# Dive into Deep Learning

ASTON ZHANG, ZACHARY C. LIPTON, MU LI, AND ALEXANDER J.
SMOLA

# Preface

Just a few years ago, there were no legions of deep learning scientists developing intelligent products and services at major companies and startups. When we entered the field, machine learning did not command headlines in daily newspapers. Our parents had no idea what machine learning was, let alone why we might prefer it to a career in medicine or law. Machine learning was a blue skies academic discipline whose industrial significance was limited to a narrow set of real-world applications, including speech recognition and computer vision. Moreover, many of these applications required so much domain knowledge that they were often regarded as entirely separate areas for which machine learning was one small component. At that time, neural networks—the predecessors of the deep learning methods that we focus on in this book—were generally regarded as outmoded.

Yet in just few years, deep learning has taken the world by surprise, driving rapid progress in such diverse fields as computer vision, natural language processing, automatic speech recognition, reinforcement learning, and biomedical informatics. Moreover, the success of deep learning in so many tasks of practical interest has even catalyzed developments in theoretical machine learning and statistics. With these advances in hand, we can now build cars that drive themselves with more autonomy than ever before (though less autonomy than some companies might have you believe), dialogue systems that debug code by asking clarifying questions, and software agents beating the best human players in the world at board games such as Go, a feat once thought to be decades away. Already, these tools exert ever-wider influence on industry and society, changing the way movies are made, diseases are diagnosed, and playing a growing role in basic sciences—from astrophysics, to climate modeling, to weather prediction, to biomedicine.

## About This Book

This book represents our attempt to make deep learning approachable, teaching you the *concepts*, the *context*, and the *code*.

### One Medium Combining Code, Math, and HTML

For any computing technology to reach its full impact, it must be well understood, well documented, and supported by mature, well-maintained tools. The key ideas should be clearly distilled, minimizing the onboarding time needed to bring new practitioners up to date. Mature libraries should automate common tasks, and exemplar code should make it easy for practitioners to modify, apply, and extend common applications to suit their needs.

As an example, take dynamic web applications. Despite a large number of companies, such as Amazon, developing successful database-driven web applications in the 1990s, the potential of this technology to aid creative entrepreneurs was realized to a far greater degree only in the past ten years, owing in part to the development of powerful, well-documented frameworks.

Testing the potential of deep learning presents unique challenges because any single application brings together various disciplines. Applying deep learning requires simultaneously understanding (i) the motivations for casting a problem in a particular way; (ii) the mathematical form of a given model; (iii) the optimization algorithms for fitting the models to data; (iv) the statistical principles that tell us when we should expect our models to generalize to unseen data and practical methods for certifying that they have, in fact, generalized; and (v) the engineering techniques required to train models efficiently, navigating the pitfalls of numerical computing and getting the most out of available hardware. Teaching the critical thinking skills required to formulate problems, the mathematics to solve them, and the software tools to implement those solutions all in one place presents formidable challenges. Our goal in this book is to present a unified resource to bring would-be practitioners up to speed.

When we started this book project, there were no resources that simultaneously (i) remained up to date; (ii) covered the breadth of modern machine learning practices with sufficient technical depth; and (iii) interleaved exposition of the quality one expects of a textbook with the clean runnable code that one expects of a hands-on tutorial. We found plenty of code examples illustrating how to use a given deep learning framework (e.g., how to do basic numerical computing with matrices in TensorFlow) or for implementing particular techniques (e.g., code snippets for LeNet, AlexNet, ResNet, etc.) scattered across various blog posts and GitHub repositories. However, these examples typically focused on *how* to implement a given approach, but left out the discussion of *why* certain algorithmic decisions are made. While some interactive resources have popped up sporadically to address a particular topic, e.g., the engaging blog posts published on the website Distill[1], or personal blogs, they only covered selected topics in deep learning, and often lacked associated code. On the other hand, while several deep learning textbooks have emerged—e.g., Goodfellow *et al.* (2016), which offers a comprehensive survey on the basics of deep learning—these resources do not marry the descriptions to realizations of the concepts in code, sometimes leaving readers clueless as to how to implement them. Moreover, too many resources are hidden behind the paywalls of commercial course providers.

1

We set out to create a resource that could (i) be freely available for everyone; (ii) offer sufficient technical depth to provide a starting point on the path to actually becoming an applied machine learning scientist; (iii) include runnable code, showing readers *how* to solve prob-lems in practice; (iv) allow for rapid updates, both by us and also by the community at large; and (v) be complemented by a forum[2] for interactive discussion of technical details and to answer questions.

2

These goals were often in conflict. Equations, theorems, and citations are best managed and laid out in LaTeX. Code is best described in Python. And webpages are native in HTML and JavaScript. Furthermore, we want the content to be accessible both as executable code, as a physical book, as a downloadable PDF, and on the Internet as a website. No workflows seemed suited to these demands, so we decided to assemble our own (Section B.6). We settled on GitHub to share the source and to facilitate community contributions; Jupyter notebooks for mixing code, equations and text; Sphinx as a rendering engine; and Discourse as a discussion platform. While our system is not perfect, these choices strike a compromise among the competing concerns. We believe that *Dive into Deep Learning* might be the first book published using such an integrated workflow.

## Learning by Doing

Many textbooks present concepts in succession, covering each in exhaustive detail. For example, the excellent textbook of Bishop (2006), teaches each topic so thoroughly that getting to the chapter on linear regression requires a nontrivial amount of work. While experts love this book precisely for its thoroughness, for true beginners, this property limits its usefulness as an introductory text.

In this book, we teach most concepts *just in time*. In other words, you will learn concepts at the very moment that they are needed to accomplish some practical end. While we take some time at the outset to teach fundamental preliminaries, like linear algebra and probability, we want you to taste the satisfaction of training your first model before worrying about more esoteric concepts.

Aside from a few preliminary notebooks that provide a crash course in the basic mathematical background, each subsequent chapter both introduces a reasonable number of new concepts and provides several self-contained working examples, using real datasets. This presented an organizational challenge. Some models might logically be grouped together in a single notebook. And some ideas might be best taught by executing several models in succession. By contrast, there is a big advantage to adhering to a policy of *one working example, one notebook*: This makes it as easy as possible for you to start your own research projects by leveraging our code. Just copy a notebook and start modifying it.

Throughout, we interleave the runnable code with background material as needed. In general, we err on the side of making tools available before explaining them fully (often filling in the background later). For instance, we might use *stochastic gradient descent* before explaining why it is useful or offering some intuition for why it works. This helps to give practitioners the necessary ammunition to solve problems quickly, at the expense of requiring the reader to trust us with some curatorial decisions.

This book teaches deep learning concepts from scratch. Sometimes, we delve into fine details about models that would typically be hidden from users by modern deep learning frameworks.

This comes up especially in the basic tutorials, where we want you to understand everything that happens in a given layer or optimizer. In these cases, we often present two versions of the example: one where we implement everything from scratch, relying only on NumPy-like functionality and automatic differentiation, and a more practical example, where we write succinct code using the high-level APIs of deep learning frameworks. After explaining how some component works, we rely on the high-level API in subsequent tutorials.

## Content and Structure

The book can be divided into roughly three parts, dealing with preliminaries, deep learning techniques, and advanced topics focused on real systems and applications (Fig. 1).

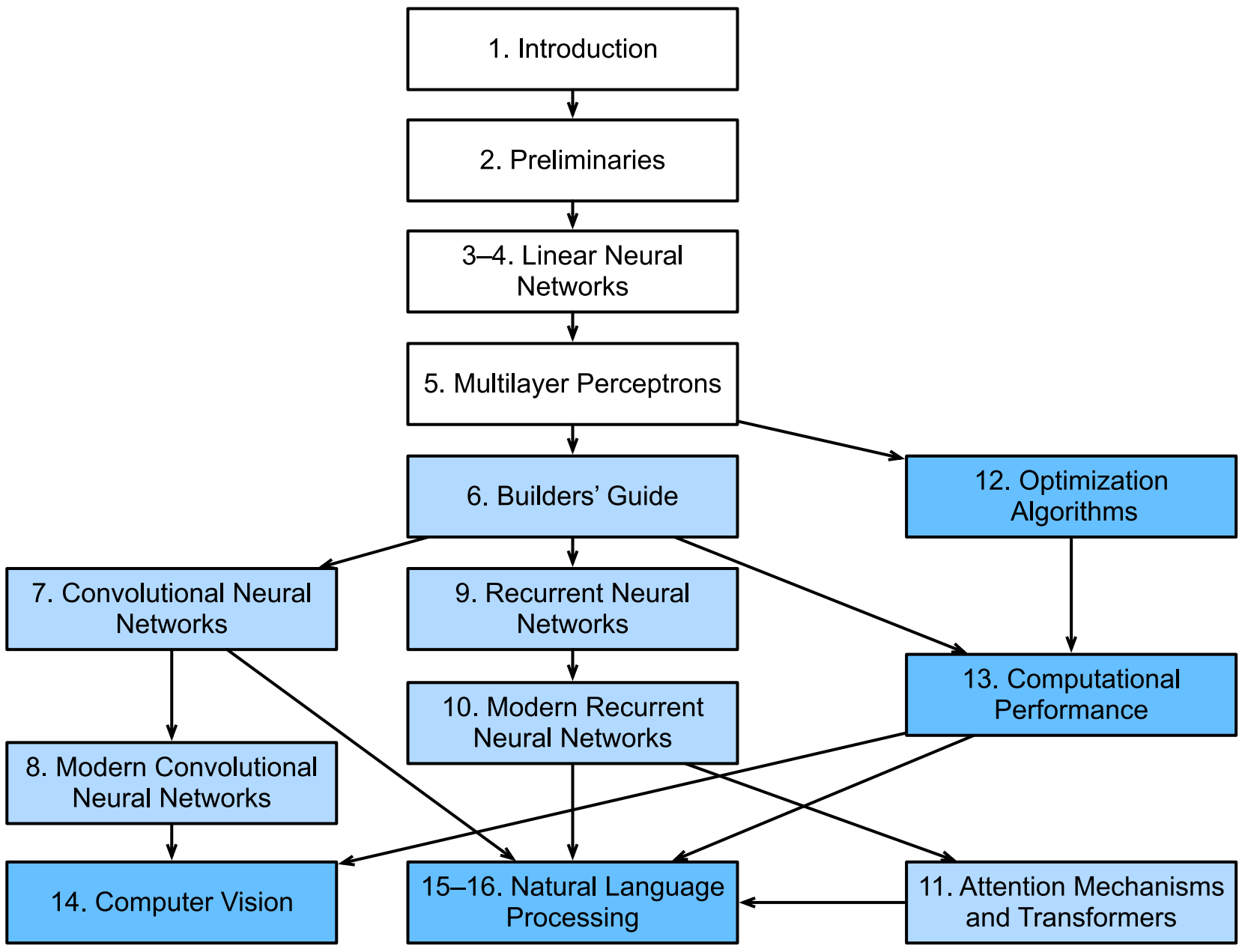

Fig. 1 Book structure.

- **Part 1: Basics and Preliminaries**. Chapter 1 is an introduction to deep learning. Then, in Chapter 2, we quickly bring you up to speed on the prerequisites required for hands-on deep learning, such as how to store and manipulate data, and how to apply various numerical operations based on elementary concepts from linear algebra, calculus, and probability. Chapter 3 and Chapter 5 cover the most fundamental concepts and techniques in deep learning, including regression and classification; linear models; multilayer perceptrons; and overfitting and regularization.

- **Part 2: Modern Deep Learning Techniques**. Chapter 6 describes the key computational components of deep learning systems and lays the groundwork for our subsequent implementations of more complex models. Next, Chapter 7 and Chapter 8 present convolutional neural networks (CNNs), powerful tools that form the backbone of most modern computer vision systems. Similarly, Chapter 9 and Chapter 10 introduce recurrent neural networks (RNNs), models that exploit sequential (e.g., temporal) structure in data and are commonly used for natural language processing and time series prediction. In Chapter 11, we describe a relatively new class of models, based on so-called *attention mechanisms*, that has displaced RNNs as the dominant architecture for most natural language processing tasks. These sections will bring you up to speed on the most powerful and general tools that are widely used by deep learning practitioners.

- **Part 3: Scalability, Efficiency, and Applications** (available online[3]). In Chapter 12, we discuss several common optimization algorithms used to train deep learning models. Next, in Chapter 13, we examine several key factors that influence the computational performance of deep learning code. Then, in Chapter 14, we illustrate major applications of deep learning in computer vision. Finally, in Chapter 15 and Chapter 16, we demonstrate how to pretrain language representation models and apply them to natural language processing tasks.

3

## Code

Most sections of this book feature executable code. We believe that some intuitions are best developed via trial and error, tweaking the code in small ways and observing the results. Ideally, an elegant mathematical theory might tell us precisely how to tweak our code to achieve a desired result. However, deep learning practitioners today must often tread where no solid theory provides guidance. Despite our best attempts, formal explanations for the efficacy of various techniques are still lacking, for a variety of reasons: the mathematics to characterize these models can be so difficult; the explanation likely depends on properties of the data that currently lack clear definitions; and serious inquiry on these topics has only recently kicked into high gear. We are hopeful that as the theory of deep learning progresses, each future edition of this book will provide insights that eclipse those presently available.

To avoid unnecessary repetition, we capture some of our most frequently imported and used functions and classes in the `d2l` package. Throughout, we mark blocks of code (such as functions, classes, or collection of import statements) with `#@save` to indicate that they will be accessed later via the `d2l` package. We offer a detailed overview of these classes and functions in Section B.8. The `d2l` package is lightweight and only requires the following dependencies:

```
#@save
import collections
import hashlib
import inspect
import math
import os
import random
import re
import shutil
import sys
import tarfile
```

```
import time
import zipfile
from collections import defaultdict
import pandas as pd
import requests
from IPython import display
from matplotlib import pyplot as plt
from matplotlib_inline import backend_inline

d2l = sys.modules[__name__]
```

Most of the code in this book is based on PyTorch, a popular open-source framework that has been enthusiastically embraced by the deep learning research community. All of the code in this book has passed tests under the latest stable version of PyTorch. However, due to the rapid development of deep learning, some code *in the print edition* may not work properly in future versions of PyTorch. We plan to keep the online version up to date. In case you encounter any problems, please consult *Installation* (page xxxiv) to update your code and runtime environment. Below lists dependencies in our PyTorch implementation.

```
#@save
import numpy as np
import torch
import torchvision
from PIL import Image
from scipy.spatial import distance_matrix
from torch import nn
from torch.nn import functional as F
from torchvision import transforms
```

## Target Audience

This book is for students (undergraduate or graduate), engineers, and researchers, who seek a solid grasp of the practical techniques of deep learning. Because we explain every concept from scratch, no previous background in deep learning or machine learning is required. Fully explaining the methods of deep learning requires some mathematics and programming, but we will only assume that you enter with some basics, including modest amounts of linear algebra, calculus, probability, and Python programming. Just in case you have forgotten anything, the online Appendix[4] provides a refresher on most of the mathematics you will find in this book. Usually, we will prioritize intuition and ideas over mathematical rigor. If you would like to extend these foundations beyond the prerequisites to understand our book, we happily recommend some other terrific resources: *Linear Analysis* by Bollobás (1999) covers linear algebra and functional analysis in great depth. *All of Statistics* (Wasserman, 2013) provides a marvelous introduction to statistics. Joe Blitzstein's books[5] and courses[6] on probability and inference are pedagogical gems. And if you have not used Python before, you may want to peruse this Python tutorial[7].

## Notebooks, Website, GitHub, and Forum

All of our notebooks are available for download on the D2L.ai website[8] and on GitHub[9]. Associated with this book, we have launched a discussion forum, located at discuss.d2l.ai[10]. Whenever you have questions on any section of the book, you can find a link to the associated discussion page at the end of each notebook.

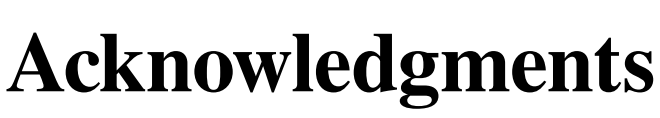

# Acknowledgments

We are indebted to the hundreds of contributors for both the English and the Chinese drafts. They helped improve the content and offered valuable feedback. This book was originally implemented with MXNet as the primary framework. We thank Anirudh Dagar and Yuan Tang for adapting a majority part of earlier MXNet code into PyTorch and TensorFlow implementations, respectively. Since July 2021, we have redesigned and reimplemented this book in PyTorch, MXNet, and TensorFlow, choosing PyTorch as the primary framework. We thank Anirudh Dagar for adapting a majority part of more recent PyTorch code into JAX implementations. We thank Gaosheng Wu, Liujun Hu, Ge Zhang, and Jiehang Xie from Baidu for adapting a majority part of more recent PyTorch code into PaddlePaddle implementations in the Chinese draft. We thank Shuai Zhang for integrating the LaTeX style from the press into the PDF building.

On GitHub, we thank every contributor of this English draft for making it better for everyone. Their GitHub IDs or names are (in no particular order): alxnorden, avinashingit, bowen0701, brettkoonce, Chaitanya Prakash Bapat, cryptonaut, Davide Fiocco, edgarroman, gkutiel, John Mitro, Liang Pu, Rahul Agarwal, Mohamed Ali Jamaoui, Michael (Stu) Stewart, Mike Müller, NRauschmayr, Prakhar Srivastav, sad-, sfermigier, Sheng Zha, sundeepteki, topecongiro, tpdi, vermicelli, Vishaal Kapoor, Vishwesh Ravi Shrimali, YaYaB, Yuhong Chen, Evgeniy Smirnov, lgov, Simon Corston-Oliver, Igor Dzreyev, Ha Nguyen, pmuens, Andrei Lukovenko, senorcinco, vfdev-5, dsweet, Mohammad Mahdi Rahimi, Abhishek Gupta, uwsd, DomKM, Lisa Oakley, Bowen Li, Aarush Ahuja, Prasanth Buddareddygari, brianhendee, mani2106, mtn, lkevinzc, caojilin, Lakshya, Fiete Lüer, Surbhi Vijayvargeeya, Muhyun Kim, dennismalmgren, adursun, Anirudh Dagar, liqingnz, Pedro Larroy, lgov, ati-ozgur, Jun Wu, Matthias Blume, Lin Yuan, geogunow, Josh Gardner, Maximilian Böther, Rakib Islam, Leonard Lausen, Abhinav Upadhyay, rongruosong, Steve Sedlmeyer, Ruslan Baratov, Rafael Schlatter, liusy182, Giannis Pappas, ati-ozgur, qbaza, dchoi77, Adam Gerson, Phuc Le, Mark Atwood, christabella, vn09, Haibin Lin, jjangga0214, RichyChen, noelo, hansent, Giel Dops, dvincent1337, WhiteD3vil, Peter Kulits, codypenta, joseppinilla, ahmaurya, karolszk, heytitle, Peter Goetz, rigtorp, Tiep Vu, sfilip, mlxd, Kale-ab Tessera, Sanjar Adilov, MatteoFerrara, hsneto, Katarzyna Biesialska, Gregory Bruss, Duy–Thanh Doan, paulaurel, graytowne, Duc Pham, sl7423, Jaedong Hwang, Yida Wang, cys4, clhm, Jean Kaddour, austinmw, trebeljahr, tbaums, Cuong V. Nguyen, pavelkomarov, vzlamal, No-

tAnotherSystem, J-Arun-Mani, jancio, eldarkurtic, the-great-shazbot, doctorcolossus, gducharme, cclauss, Daniel-Mietchen, hoonose, biagiom, abhinavsp0730, jonathanhrandall, ysraell, Nodar Okroshiashvili, UgurKap, Jiyang Kang, StevenJokes, Tomer Kaftan, liweiwp, netyster, ypandya, NishantTharani, heiligerl, SportsTHU, Hoa Nguyen, manuel-arno-korfmann-webentwicklung, aterzis-personal, nxby, Xiaoting He, Josiah Yoder, mathresearch, mzz2017, jroberayalas, iluu, ghejc, BSharmi, vkramdev, simonwardjones, LakshKD, TalNeoran, djliden, Nikhil95, Oren Barkan, guoweis, haozhu233, pratikhack, Yue Ying, tayfununal, steinsag, charleybeller, Andrew Lumsdaine, Jiekui Zhang, Deepak Pathak, Florian Donhauser, Tim Gates, Adriaan Tijsseling, Ron Medina, Gaurav Saha, Murat Semerci, Lei Mao, Levi McClenny, Joshua Broyde, jake221, jonbally, zyhazwraith, Brian Pulfer, Nick Tomasino, Lefan Zhang, Hongshen Yang, Vinney Cavallo, yuntai, Yuanxiang Zhu, amarazov, pasricha, Ben Greenawald, Shivam Upadhyay, Quanshangze Du, Biswajit Sahoo, Parthe Pandit, Ishan Kumar, HomunculusK, Lane Schwartz, varadgunjal, Jason Wiener, Armin Gholampoor, Shreshtha13, eigenarnav, Hyeonggyu Kim, EmilyOng, Bálint Mucsányi, Chase DuBois, Juntian Tao, Wenxiang Xu, Lifu Huang, filevich, quake2005, nils-werner, Yiming Li, Marsel Khisamutdinov, Francesco "Fuma" Fumagalli, Peilin Sun, Vincent Gurgul, qingfengtommy, Janmey Shukla, Mo Shan, Kaan Sancak, regob, AlexSauer, Gopalakrishna Ramachandra, Tobias Uelwer, Chao Wang, Tian Cao, Nicolas Corthorn, akash5474, kxxt, zxydi1992, Jacob Britton, Shuangchi He, zhmou, krahets, Jie-Han Chen, Atishay Garg, Marcel Flygare, adtygan, Nik Vaessen, bolded, Louis Schlessinger, Balaji Varatharajan, atgctg, Kaixin Li, Victor Barbaros, Riccardo Musto, Elizabeth Ho, azimjonn, Guilherme Miotto, Alessandro Finamore, Joji Joseph, Anthony Biel, Zeming Zhao, shjustinbaek, gab-chen, nantekoto, Yutaro Nishiyama, Oren Amsalem, Tian-MaoMao, Amin Allahyar, Gijs van Tulder, Mikhail Berkov, iamorphen, Matthew Caseres, Andrew Walsh, pggPL, RohanKarthikeyan, Ryan Choi, and Likun Lei.

We thank Amazon Web Services, especially Wen-Ming Ye, George Karypis, Swami Sivasubramanian, Peter DeSantis, Adam Selipsky, and Andrew Jassy for their generous support in writing this book. Without the available time, resources, discussions with colleagues, and continuous encouragement, this book would not have happened. During the preparation of the book for publication, Cambridge University Press has offered excellent support. We thank our commissioning editor David Tranah for his help and professionalism.

# Summary

Deep learning has revolutionized pattern recognition, introducing technology that now powers a wide range of technologies, in such diverse fields as computer vision, natural language processing, and automatic speech recognition. To successfully apply deep learning, you must understand how to cast a problem, the basic mathematics of modeling, the algorithms for fitting your models to data, and the engineering techniques to implement it all. This book presents a comprehensive resource, including prose, figures, mathematics, and code, all in one place.

## Exercises

1. Register an account on the discussion forum of this book discuss.d2l.ai[11].

11

2. Install Python on your computer.
3. Follow the links at the bottom of the section to the forum, where you will be able to seek out help and discuss the book and find answers to your questions by engaging the authors and broader community.

Discussions[12].

# Installation

In order to get up and running, we will need an environment for running Python, the Jupyter Notebook, the relevant libraries, and the code needed to run the book itself.

## Installing Miniconda

13

Your simplest option is to install Miniconda[13]. Note that the Python 3.x version is required. You can skip the following steps if your machine already has conda installed.

Visit the Miniconda website and determine the appropriate version for your system based on your Python 3.x version and machine architecture. Suppose that your Python version is 3.9 (our tested version). If you are using macOS, you would download the bash script whose name contains the strings "MacOSX", navigate to the download location, and execute the installation as follows (taking Intel Macs as an example):

```
# The file name is subject to changes
sh Miniconda3-py39_4.12.0-MacOSX-x86_64.sh -b
```

A Linux user would download the file whose name contains the strings "Linux" and execute the following at the download location:

```
# The file name is subject to changes
sh Miniconda3-py39_4.12.0-Linux-x86_64.sh -b
```

14

A Windows user would download and install Miniconda by following its online instructions [14]. On Windows, you may search for `cmd` to open the Command Prompt (command-line interpreter) for running commands.

Next, initialize the shell so we can run `conda` directly.

```
~/miniconda3/bin/conda init
```

Then close and reopen your current shell. You should be able to create a new environment as follows:

```
conda create --name d2l python=3.9 -y
```

Now we can activate the d2l environment:

```
conda activate d2l
```

## Installing the Deep Learning Framework and the d2l Package

Before installing any deep learning framework, please first check whether or not you have proper GPUs on your machine (the GPUs that power the display on a standard laptop are not relevant for our purposes). For example, if your computer has NVIDIA GPUs and has installed CUDA[15], then you are all set. If your machine does not house any GPU, there is no need to worry just yet. Your CPU provides more than enough horsepower to get you through the first few chapters. Just remember that you will want to access GPUs before running larger models.

15

You can install PyTorch (the specified versions are tested at the time of writing) with either CPU or GPU support as follows:

```
pip install torch==2.0.0 torchvision==0.15.1
```

Our next step is to install the d2l package that we developed in order to encapsulate frequently used functions and classes found throughout this book:

```
pip install d2l==1.0.3
```

## Downloading and Running the Code

Next, you will want to download the notebooks so that you can run each of the book's code blocks. Simply click on the "Notebooks" tab at the top of any HTML page on the D2L.ai website[16] to download the code and then unzip it. Alternatively, you can fetch the notebooks from the command line as follows:

16

```
mkdir d2l-en && cd d2l-en
curl https://d2l.ai/d2l-en-1.0.3.zip -o d2l-en.zip
unzip d2l-en.zip && rm d2l-en.zip
cd pytorch
```

If you do not already have `unzip` installed, first run `sudo apt-get install unzip`. Now we can start the Jupyter Notebook server by running:

```
jupyter notebook
```

At this point, you can open http://localhost:8888 (it may have already opened automatically) in your web browser. Then we can run the code for each section of the book. Whenever you open a new command line window, you will need to execute `conda activate d2l` to activate the runtime environment before running the D2L notebooks, or updating your packages (either the deep learning framework or the `d2l` package). To exit the environment, run `conda deactivate`.

Discussions[17].

17

# Notation

Throughout this book, we adhere to the following notational conventions. Note that some of these symbols are placeholders, while others refer to specific objects. As a general rule of thumb, the indefinite article "a" often indicates that the symbol is a placeholder and that similarly formatted symbols can denote other objects of the same type. For example, "$x$: a scalar" means that lowercased letters generally represent scalar values, but "$\mathbb{Z}$: the set of integers" refers specifically to the symbol $\mathbb{Z}$.

## Numerical Objects

- $x$: a scalar
- $\mathbf{x}$: a vector
- $\mathbf{X}$: a matrix
- $\mathsf{X}$: a general tensor
- $\mathbf{I}$: the identity matrix (of some given dimension), i.e., a square matrix with 1 on all diagonal entries and 0 on all off-diagonals
- $x_i$, $[\mathbf{x}]_i$: the $i^\textrm{th}$ element of vector $\mathbf{x}$
- $x_{ij}$, $x_{i,j}$,$[\mathbf{X}]_{ij}$, $[\mathbf{X}]_{i,j}$: the element of matrix $\mathbf{X}$ at row $i$ and column $j$.

## Set Theory

- $\mathcal{X}$: a set
- $\mathbb{Z}$: the set of integers
- $\mathbb{Z}^+$: the set of positive integers

- $\mathbb{R}$: the set of real numbers
- $\mathbb{R}^n$: the set of $n$-dimensional vectors of real numbers
- $\mathbb{R}^{a\times b}$: The set of matrices of real numbers with $a$ rows and $b$ columns
- $|\mathcal{X}|$: cardinality (number of elements) of set $\mathcal{X}$
- $\mathcal{A} \cup \mathcal{B}$: union of sets $\mathcal{A}$ and $\mathcal{B}$
- $\mathcal{A} \cap \mathcal{B}$: intersection of sets $\mathcal{A}$ and $\mathcal{B}$
- $\mathcal{A} \setminus \mathcal{B}$: set subtraction of $\mathcal{B}$ from $\mathcal{A}$ (contains only those elements of $\mathcal{A}$ that do not belong to $\mathcal{B}$)

# Functions and Operators

- $f(\cdot)$: a function
- $\log(\cdot)$: the natural logarithm (base $e$)
- $\log_2(\cdot)$: logarithm to base $2$
- $\exp(\cdot)$: the exponential function
- $\mathbf{1}(\cdot)$: the indicator function; evaluates to $1$ if the boolean argument is true, and $0$ otherwise
- $\mathbf{1}_{\mathcal{X}}(z)$: the set-membership indicator function; evaluates to $1$ if the element $z$ belongs to the set $\mathcal{X}$ and $0$ otherwise
- $(\cdot)^\top$: transpose of a vector or a matrix
- $\mathbf{X}^{-1}$: inverse of matrix $\mathbf{X}$
- $\odot$: Hadamard (elementwise) product
- $[\cdot, \cdot]$: concatenation
- $\|\cdot\|_p$: $\ell_p$ norm
- $\|\cdot\|$: $\ell_2$ norm
- $\langle \mathbf{x}, \mathbf{y} \rangle$: inner (dot) product of vectors $\mathbf{x}$ and $\mathbf{y}$
- $\sum$: summation over a collection of elements
- $\prod$: product over a collection of elements
- $\stackrel{\mathrm{def}}{=}$: an equality asserted as a definition of the symbol on the left-hand side

# Calculus

- $\frac{dy}{dx}$: derivative of $y$ with respect to $x$
- $\frac{\partial y}{\partial x}$: partial derivative of $y$ with respect to $x$
- $\nabla_{\mathbf{x}} y$: gradient of $y$ with respect to $\mathbf{x}$
- $\int_a^b f(x)\;dx$: definite integral of $f$ from $a$ to $b$ with respect to $x$
- $\int f(x)\;dx$: indefinite integral of $f$ with respect to $x$

# Probability and Information Theory

- $X$: a random variable
- $P$: a probability distribution
- $X \sim P$: the random variable $X$ follows distribution $P$
- $P(X = x)$: the probability assigned to the event where random variable $X$ takes value $x$
- $P(X \mid Y)$: the conditional probability distribution of $X$ given $Y$
- $p(\cdot)$: a probability density function (PDF) associated with distribution $P$
- $E[X]$: expectation of a random variable $X$
- $X \perp Y$: random variables $X$ and $Y$ are independent
- $X \perp Y \mid Z$: random variables $X$ and $Y$ are conditionally independent given $Z$
- $\sigma_X$: standard deviation of random variable $X$
- $\mathrm{Var}(X)$: variance of random variable $X$, equal to $\sigma^2_X$
- $\mathrm{Cov}(X, Y)$: covariance of random variables $X$ and $Y$
- $\rho(X, Y)$: the Pearson correlation coefficient between $X$ and $Y$, equals $\frac{\mathrm{Cov}(X,Y)}{\sigma_X \sigma_Y}$
- $H(X)$: entropy of random variable $X$
- $D_{\mathrm{KL}}(P\|Q)$: the KL-divergence (or relative entropy) from distribution $Q$ to distribution $P$

Discussions[18].

18

# 1 Introduction

Until recently, nearly every computer program that you might have interacted with during an ordinary day was coded up as a rigid set of rules specifying precisely how it should behave. Say that we wanted to write an application to manage an e-commerce platform. After huddling around a whiteboard for a few hours to ponder the problem, we might settle on the broad strokes of a working solution, for example: (i) users interact with the application through an interface running in a web browser or mobile application; (ii) our application interacts with a commercial-grade database engine to keep track of each user's state and maintain records of historical transactions; and (iii) at the heart of our application, the *business logic* (you might say, the *brains*) of our application spells out a set of rules that map every conceivable circumstance to the corresponding action that our program should take.

To build the brains of our application, we might enumerate all the common events that our program should handle. For example, whenever a customer clicks to add an item to their shopping cart, our program should add an entry to the shopping cart database table, associating that user's ID with the requested product's ID. We might then attempt to step through every possible corner case, testing the appropriateness of our rules and making any necessary modifications. What happens if a user initiates a purchase with an empty cart? While few developers ever get it completely right the first time (it might take some test runs to work out the kinks), for the most part we can write such programs and confidently launch them *before* ever seeing a real customer. Our ability to manually design automated systems that drive functioning products and systems, often in novel situations, is a remarkable cognitive feat. And when you are able to devise solutions that work $100\%$ of the time, you typically should not be worrying about machine learning.

Fortunately for the growing community of machine learning scientists, many tasks that we would like to automate do not bend so easily to human ingenuity. Imagine huddling around the whiteboard with the smartest minds you know, but this time you are tackling one of the following problems:

- Write a program that predicts tomorrow's weather given geographic information, satellite images, and a trailing window of past weather.
- Write a program that takes in a factoid question, expressed in free-form text, and answers it correctly.
- Write a program that, given an image, identifies every person depicted in it and draws outlines around each.

- Write a program that presents users with products that they are likely to enjoy but unlikely, in the natural course of browsing, to encounter.

For these problems, even elite programmers would struggle to code up solutions from scratch. The reasons can vary. Sometimes the program that we are looking for follows a pattern that changes over time, so there is no fixed right answer! In such cases, any successful solution must adapt gracefully to a changing world. At other times, the relationship (say between pixels, and abstract categories) may be too complicated, requiring thousands or millions of computations and following unknown principles. In the case of image recognition, the precise steps required to perform the task lie beyond our conscious understanding, even though our subconscious cognitive processes execute the task effortlessly.

*Machine learning* is the study of algorithms that can learn from experience. As a machine learning algorithm accumulates more experience, typically in the form of observational data or interactions with an environment, its performance improves. Contrast this with our deterministic e-commerce platform, which follows the same business logic, no matter how much experience accrues, until the developers themselves learn and decide that it is time to update the software. In this book, we will teach you the fundamentals of machine learning, focusing in particular on *deep learning*, a powerful set of techniques driving innovations in areas as diverse as computer vision, natural language processing, healthcare, and genomics.

## 1.1 A Motivating Example

Before beginning writing, the authors of this book, like much of the work force, had to become caffeinated. We hopped in the car and started driving. Using an iPhone, Alex called out “Hey Siri”, awakening the phone’s voice recognition system. Then Mu commanded “directions to Blue Bottle coffee shop”. The phone quickly displayed the transcription of his command. It also recognized that we were asking for directions and launched the Maps application (app) to fulfill our request. Once launched, the Maps app identified a number of routes. Next to each route, the phone displayed a predicted transit time. While this story was fabricated for pedagogical convenience, it demonstrates that in the span of just a few seconds, our everyday interactions with a smart phone can engage several machine learning models.

Imagine just writing a program to respond to a *wake word* such as “Alexa”, “OK Google”, and “Hey Siri”. Try coding it up in a room by yourself with nothing but a computer and a code editor, as illustrated in Fig. 1.1.1. How would you write such a program from first principles? Think about it… the problem is hard. Every second, the microphone will collect roughly 44,000 samples. Each sample is a measurement of the amplitude of the sound wave. What rule could map reliably from a snippet of raw audio to confident predictions {yes, no} about whether the snippet contains the wake word? If you are stuck, do not worry. We do not know how to write such a program from scratch either. That is why we use machine learning.

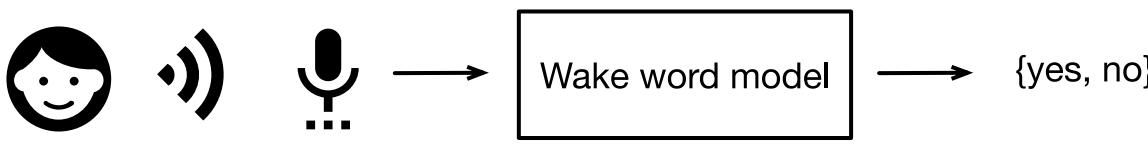

Fig. 1.1.1 Identify a wake word.

Here is the trick. Often, even when we do not know how to tell a computer explicitly how to map from inputs to outputs, we are nonetheless capable of performing the cognitive feat ourselves. In other words, even if you do not know how to program a computer to recognize the word "Alexa", you yourself are able to recognize it. Armed with this ability, we can collect a huge *dataset* containing examples of audio snippets and associated labels, indicating which snippets contain the wake word. In the currently dominant approach to machine learning, we do not attempt to design a system *explicitly* to recognize wake words. Instead, we define a flexible program whose behavior is determined by a number of *parameters*. Then we use the dataset to determine the best possible parameter values, i.e., those that improve the performance of our program with respect to a chosen performance measure.

You can think of the parameters as knobs that we can turn, manipulating the behavior of the program. Once the parameters are fixed, we call the program a *model*. The set of all distinct programs (input–output mappings) that we can produce just by manipulating the parameters is called a *family* of models. And the "meta-program" that uses our dataset to choose the parameters is called a *learning algorithm*.

Before we can go ahead and engage the learning algorithm, we have to define the problem precisely, pinning down the exact nature of the inputs and outputs, and choosing an appropriate model family. In this case, our model receives a snippet of audio as *input*, and the model generates a selection among $\{\text{yes}, \text{no}\}$ as *output*. If all goes according to plan the model's guesses will typically be correct as to whether the snippet contains the wake word.

If we choose the right family of models, there should exist one setting of the knobs such that the model fires "yes" every time it hears the word "Alexa". Because the exact choice of the wake word is arbitrary, we will probably need a model family sufficiently rich that, via another setting of the knobs, it could fire "yes" only upon hearing the word "Apricot". We expect that the same model family should be suitable for "Alexa" recognition and "Apricot" recognition because they seem, intuitively, to be similar tasks. However, we might need a different family of models entirely if we want to deal with fundamentally different inputs or outputs, say if we wanted to map from images to captions, or from English sentences to Chinese sentences.

As you might guess, if we just set all of the knobs randomly, it is unlikely that our model will recognize "Alexa", "Apricot", or any other English word. In machine learning, the *learning* is the process by which we discover the right setting of the knobs for coercing the desired behavior from our model. In other words, we *train* our model with data. As shown in Fig. 1.1.2, the training process usually looks like the following:

1. Start off with a randomly initialized model that cannot do anything useful.
2. Grab some of your data (e.g., audio snippets and corresponding {yes, no} labels).
3. Tweak the knobs to make the model perform better as assessed on those examples.
4. Repeat Steps 2 and 3 until the model is awesome.

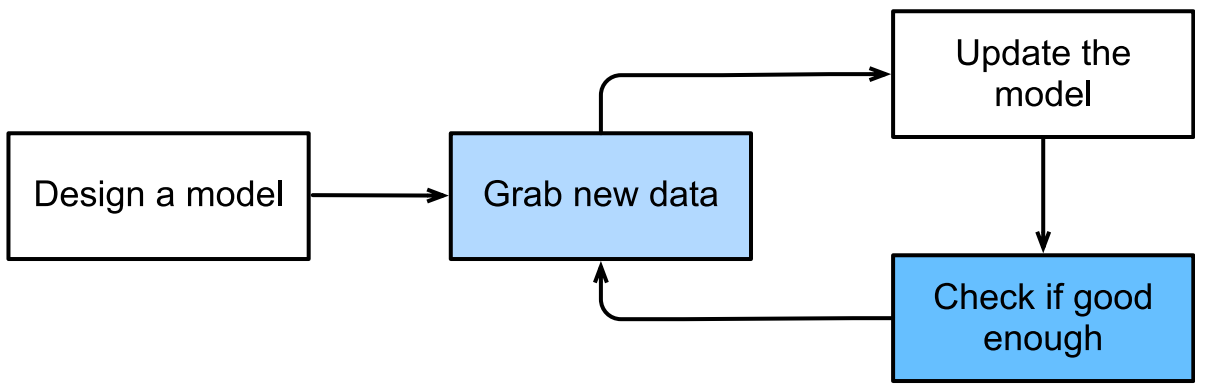

Fig. 1.1.2 A typical training process.

To summarize, rather than code up a wake word recognizer, we code up a program that can *learn* to recognize wake words, if presented with a large labeled dataset. You can think of this act of determining a program's behavior by presenting it with a dataset as *programming with data*. That is to say, we can "program" a cat detector by providing our machine learning system with many examples of cats and dogs. This way the detector will eventually learn to emit a very large positive number if it is a cat, a very large negative number if it is a dog, and something closer to zero if it is not sure. This barely scratches the surface of what machine learning can do. Deep learning, which we will explain in greater detail later, is just one among many popular methods for solving machine learning problems.

## 1.2 Key Components

In our wake word example, we described a dataset consisting of audio snippets and binary labels, and we gave a hand-wavy sense of how we might train a model to approximate a mapping from snippets to classifications. This sort of problem, where we try to predict a designated unknown label based on known inputs given a dataset consisting of examples for which the labels are known, is called *supervised learning*. This is just one among many kinds of machine learning problems. Before we explore other varieties, we would like to shed more light on some core components that will follow us around, no matter what kind of machine learning problem we tackle:

1. The *data* that we can learn from.
2. A *model* of how to transform the data.
3. An *objective function* that quantifies how well (or badly) the model is doing.
4. An *algorithm* to adjust the model's parameters to optimize the objective function.

### 1.2.1 Data

It might go without saying that you cannot do data science without data. We could lose hundreds of pages pondering what precisely data *is*, but for now, we will focus on the key properties of the datasets that we will be concerned with. Generally, we are concerned with a collection of examples. In order to work with data usefully, we typically need to come up with a suitable numerical representation. Each *example* (or *data point*, *data instance*, *sample*) typically consists of a set of attributes called *features* (sometimes called *covariates* or *inputs*), based on which the model must make its predictions. In supervised learning problems, our goal is to predict the value of a special attribute, called the *label* (or *target*), that is not part of the model's input.

If we were working with image data, each example might consist of an individual photograph (the features) and a number indicating the category to which the photograph belongs (the label). The photograph would be represented numerically as three grids of numerical values representing the brightness of red, green, and blue light at each pixel location. For example, a $200 \times 200$ pixel color photograph would consist of $200 \times 200 \times 3 = 120000$ numerical values.

Alternatively, we might work with electronic health record data and tackle the task of predicting the likelihood that a given patient will survive the next 30 days. Here, our features might consist of a collection of readily available attributes and frequently recorded measurements, including age, vital signs, comorbidities, current medications, and recent procedures. The label available for training would be a binary value indicating whether each patient in the historical data survived within the 30-day window.

In such cases, when every example is characterized by the same number of numerical features, we say that the inputs are fixed-length vectors and we call the (constant) length of the vectors the *dimensionality* of the data. As you might imagine, fixed-length inputs can be convenient, giving us one less complication to worry about. However, not all data can easily be represented as *fixed-length* vectors. While we might expect microscope images to come from standard equipment, we cannot expect images mined from the Internet all to have the same resolution or shape. For images, we might consider cropping them to a standard size, but that strategy only gets us so far. We risk losing information in the cropped-out portions. Moreover, text data resists fixed-length representations even more stubbornly. Consider the customer reviews left on e-commerce sites such as Amazon, IMDb, and TripAdvisor. Some are short: "it stinks!". Others ramble for pages. One major advantage of deep learning over traditional methods is the comparative grace with which modern models can handle *varying-length* data.

Generally, the more data we have, the easier our job becomes. When we have more data, we can train more powerful models and rely less heavily on preconceived assumptions. The regime change from (comparatively) small to big data is a major contributor to the success of modern deep learning. To drive the point home, many of the most exciting models in deep learning do not work without large datasets. Some others might work in the small data regime, but are no better than traditional approaches.

Finally, it is not enough to have lots of data and to process it cleverly. We need the *right* data.

If the data is full of mistakes, or if the chosen features are not predictive of the target quantity of interest, learning is going to fail. The situation is captured well by the cliché: *garbage in, garbage out*. Moreover, poor predictive performance is not the only potential consequence. In sensitive applications of machine learning, like predictive policing, resume screening, and risk models used for lending, we must be especially alert to the consequences of garbage data. One commonly occurring failure mode concerns datasets where some groups of people are unrepresented in the training data. Imagine applying a skin cancer recognition system that had never seen black skin before. Failure can also occur when the data does not only under-represent some groups but reflects societal prejudices. For example, if past hiring decisions are used to train a predictive model that will be used to screen resumes then machine learning models could inadvertently capture and automate historical injustices. Note that this can all happen without the data scientist actively conspiring, or even being aware.

### 1.2.2 Models

Most machine learning involves transforming the data in some sense. We might want to build a system that ingests photos and predicts smiley-ness. Alternatively, we might want to ingest a set of sensor readings and predict how normal vs. anomalous the readings are. By *model*, we denote the computational machinery for ingesting data of one type, and spitting out predictions of a possibly different type. In particular, we are interested in *statistical models* that can be estimated from data. While simple models are perfectly capable of addressing appropriately simple problems, the problems that we focus on in this book stretch the limits of classical methods. Deep learning is differentiated from classical approaches principally by the set of powerful models that it focuses on. These models consist of many successive transformations of the data that are chained together top to bottom, thus the name *deep learning*. On our way to discussing deep models, we will also discuss some more traditional methods.

### 1.2.3 Objective Functions

Earlier, we introduced machine learning as learning from experience. By *learning* here, we mean improving at some task over time. But who is to say what constitutes an improvement? You might imagine that we could propose updating our model, and some people might disagree on whether our proposal constituted an improvement or not.

In order to develop a formal mathematical system of learning machines, we need to have formal measures of how good (or bad) our models are. In machine learning, and optimization more generally, we call these *objective functions*. By convention, we usually define objective functions so that lower is better. This is merely a convention. You can take any function for which higher is better, and turn it into a new function that is qualitatively identical but for which lower is better by flipping the sign. Because we choose lower to be better, these functions are sometimes called *loss functions*.

When trying to predict numerical values, the most common loss function is *squared error*, i.e., the square of the difference between the prediction and the ground truth target. For classification, the most common objective is to minimize error rate, i.e., the fraction of examples on which our predictions disagree with the ground truth. Some objectives (e.g., squared error) are easy to optimize, while others (e.g., error rate) are difficult to optimize directly, owing to non-differentiability or other complications. In these cases, it is common instead to optimize a *surrogate objective*.

During optimization, we think of the loss as a function of the model's parameters, and treat the training dataset as a constant. We learn the best values of our model's parameters by minimizing the loss incurred on a set consisting of some number of examples collected for training. However, doing well on the training data does not guarantee that we will do well on unseen data. So we will typically want to split the available data into two partitions: the *training dataset* (or *training set*), for learning model parameters; and the *test dataset* (or *test set*), which is held out for evaluation. At the end of the day, we typically report how our models perform on both partitions. You could think of training performance as analogous to the scores that a student achieves on the practice exams used to prepare for some real final exam. Even if the results are encouraging, that does not guarantee success on the final exam. Over the course of studying, the student might begin to memorize the practice questions, appearing to master the topic but faltering when faced with previously unseen questions on the actual final exam. When a model performs well on the training set but fails to generalize to unseen data, we say that it is *overfitting* to the training data.

### 1.2.4 Optimization Algorithms

Once we have got some data source and representation, a model, and a well-defined objective function, we need an algorithm capable of searching for the best possible parameters for minimizing the loss function. Popular optimization algorithms for deep learning are based on an approach called *gradient descent*. In brief, at each step, this method checks to see, for each parameter, how that training set loss would change if you perturbed that parameter by just a small amount. It would then update the parameter in the direction that lowers the loss.

## 1.3 Kinds of Machine Learning Problems

The wake word problem in our motivating example is just one among many that machine learning can tackle. To motivate the reader further and provide us with some common language that will follow us throughout the book, we now provide a broad overview of the landscape of machine learning problems.

### 1.3.1 Supervised Learning

Supervised learning describes tasks where we are given a dataset containing both features and labels and asked to produce a model that predicts the labels when given input features.

Each feature–label pair is called an example. Sometimes, when the context is clear, we may use the term *examples* to refer to a collection of inputs, even when the corresponding labels are unknown. The supervision comes into play because, for choosing the parameters, we (the supervisors) provide the model with a dataset consisting of labeled examples. In probabilistic terms, we typically are interested in estimating the conditional probability of a label given input features. While it is just one among several paradigms, supervised learning accounts for the majority of successful applications of machine learning in industry. Partly that is because many important tasks can be described crisply as estimating the probability of something unknown given a particular set of available data:

- Predict cancer vs. not cancer, given a computer tomography image.
- Predict the correct translation in French, given a sentence in English.
- Predict the price of a stock next month based on this month's financial reporting data.

While all supervised learning problems are captured by the simple description "predicting the labels given input features", supervised learning itself can take diverse forms and require tons of modeling decisions, depending on (among other considerations) the type, size, and quantity of the inputs and outputs. For example, we use different models for processing sequences of arbitrary lengths and fixed-length vector representations. We will visit many of these problems in depth throughout this book.

Informally, the learning process looks something like the following. First, grab a big collection of examples for which the features are known and select from them a random subset, acquiring the ground truth labels for each. Sometimes these labels might be available data that have already been collected (e.g., did a patient die within the following year?) and other times we might need to employ human annotators to label the data, (e.g., assigning images to categories). Together, these inputs and corresponding labels comprise the training set. We feed the training dataset into a supervised learning algorithm, a function that takes as input a dataset and outputs another function: the learned model. Finally, we can feed previously unseen inputs to the learned model, using its outputs as predictions of the corresponding label. The full process is drawn in Fig. 1.3.1.

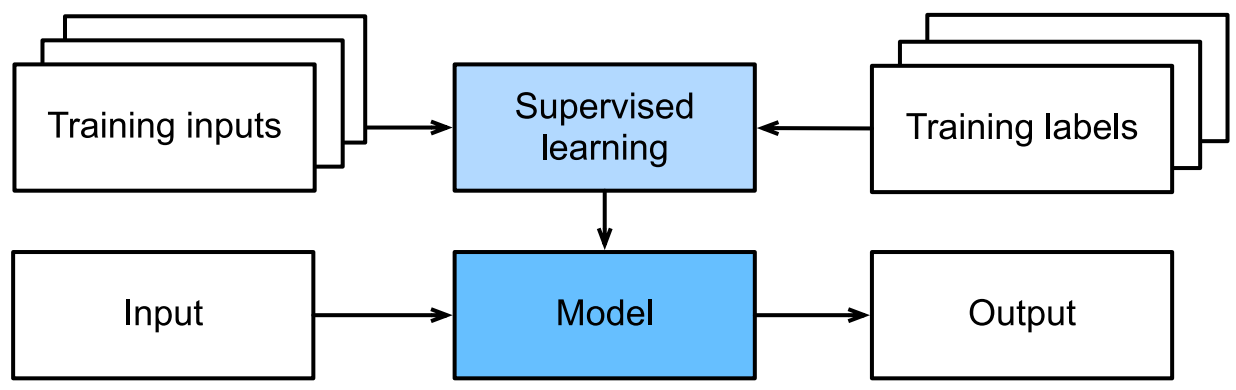

Fig. 1.3.1 Supervised learning.

### Regression

Perhaps the simplest supervised learning task to wrap your head around is *regression*. Consider, for example, a set of data harvested from a database of home sales. We might construct a table, in which each row corresponds to a different house, and each column corresponds to some relevant attribute, such as the square footage of a house, the number of bedrooms, the number of bathrooms, and the number of minutes (walking) to the center of town. In this dataset, each example would be a specific house, and the corresponding feature vector would be one row in the table. If you live in New York or San Francisco, and you are not the CEO of Amazon, Google, Microsoft, or Facebook, the (sq. footage, no. of bedrooms, no. of bathrooms, walking distance) feature vector for your home might look something like: $[600, 1, 1, 60]$. However, if you live in Pittsburgh, it might look more like $[3000, 4, 3, 10]$. Fixed-length feature vectors like this are essential for most classic machine learning algorithms.

What makes a problem a regression is actually the form of the target. Say that you are in the market for a new home. You might want to estimate the fair market value of a house, given some features such as above. The data here might consist of historical home listings and the labels might be the observed sales prices. When labels take on arbitrary numerical values (even within some interval), we call this a *regression* problem. The goal is to produce a model whose predictions closely approximate the actual label values.

Lots of practical problems are easily described as regression problems. Predicting the rating that a user will assign to a movie can be thought of as a regression problem and if you designed a great algorithm to accomplish this feat in 2009, you might have won the 1-million-dollar Netflix prize[19]. Predicting the length of stay for patients in the hospital is also a regression problem. A good rule of thumb is that any *how much?* or *how many?* problem is likely to be regression. For example:

- How many hours will this surgery take?
- How much rainfall will this town have in the next six hours?

Even if you have never worked with machine learning before, you have probably worked through a regression problem informally. Imagine, for example, that you had your drains repaired and that your contractor spent 3 hours removing gunk from your sewage pipes. Then they sent you a bill of 350 dollars. Now imagine that your friend hired the same contractor for 2 hours and received a bill of 250 dollars. If someone then asked you how much to expect on their upcoming gunk-removal invoice you might make some reasonable assumptions, such as more hours worked costs more dollars. You might also assume that there is some base charge and that the contractor then charges per hour. If these assumptions held true, then given these two data examples, you could already identify the contractor's pricing structure: 100 dollars per hour plus 50 dollars to show up at your house. If you followed that much, then you already understand the high-level idea behind *linear* regression.

In this case, we could produce the parameters that exactly matched the contractor's prices. Sometimes this is not possible, e.g., if some of the variation arises from factors beyond your two features. In these cases, we will try to learn models that minimize the distance between our predictions and the observed values. In most of our chapters, we will focus on minimizing the squared error loss function. As we will see later, this loss corresponds to the assumption that our data were corrupted by Gaussian noise.

### Classification

While regression models are great for addressing *how many?* questions, lots of problems do not fit comfortably in this template. Consider, for example, a bank that wants to develop a check scanning feature for its mobile app. Ideally, the customer would simply snap a photo of a check and the app would automatically recognize the text from the image. Assuming that we had some ability to segment out image patches corresponding to each handwritten character, then the primary remaining task would be to determine which character among some known set is depicted in each image patch. These kinds of *which one?* problems are called *classification* and require a different set of tools from those used for regression, although many techniques will carry over.

In *classification*, we want our model to look at features, e.g., the pixel values in an image, and then predict to which *category* (sometimes called a *class*) among some discrete set of options, an example belongs. For handwritten digits, we might have ten classes, corresponding to the digits 0 through 9. The simplest form of classification is when there are only two classes, a problem which we call *binary classification*. For example, our dataset could consist of images of animals and our labels might be the classes $\{\text{cat}, \text{dog}\}$. Whereas in regression we sought a regressor to output a numerical value, in classification we seek a classifier, whose output is the predicted class assignment.

For reasons that we will get into as the book gets more technical, it can be difficult to optimize a model that can only output a *firm* categorical assignment, e.g., either "cat" or "dog". In these cases, it is usually much easier to express our model in the language of probabilities. Given features of an example, our model assigns a probability to each possible class. Returning to our animal classification example where the classes are $\{\text{cat}, \text{dog}\}$, a classifier might see an image and output the probability that the image is a cat as 0.9. We can interpret this number by saying that the classifier is 90% sure that the image depicts a cat. The magnitude of the probability for the predicted class conveys a notion of uncertainty. It is not the only one available and we will discuss others in chapters dealing with more advanced topics.

When we have more than two possible classes, we call the problem *multiclass classification*. Common examples include handwritten character recognition $\{0, 1, 2, ... 9, \mathrm{a}, \mathrm{b}, \mathrm{c}, ...\}$. While we attacked regression problems by trying to minimize the squared error loss function, the common loss function for classification problems is called *cross-entropy*, whose name will be demystified when we introduce information theory in later chapters.

Note that the most likely class is not necessarily the one that you are going to use for your decision. Assume that you find a beautiful mushroom in your backyard as shown in Fig. 1.3.2.

Now, assume that you built a classifier and trained it to predict whether a mushroom is poisonous based on a photograph. Say our poison-detection classifier outputs that the probability that Fig. 1.3.2 shows a death cap is 0.2. In other words, the classifier is 80% sure that our mushroom is not a death cap. Still, you would have to be a fool to eat it. That is because the certain benefit of a delicious dinner is not worth a 20% risk of dying from it. In other words, the effect of the uncertain risk outweighs the benefit by far. Thus, in order to make a decision about whether to eat the mushroom, we need to compute the expected detriment associated with each action which depends both on the likely outcomes and the benefits or harms associated with each. In this case, the detriment incurred by eating the mushroom might be $0.2 \times \infty + 0.8 \times 0 = \infty$, whereas the loss of discarding it is $0.2 \times 0 + 0.8 \times 1 = 0.8$. Our caution was justified: as any mycologist would tell us, the mushroom in Fig. 1.3.2 is actually a death cap.

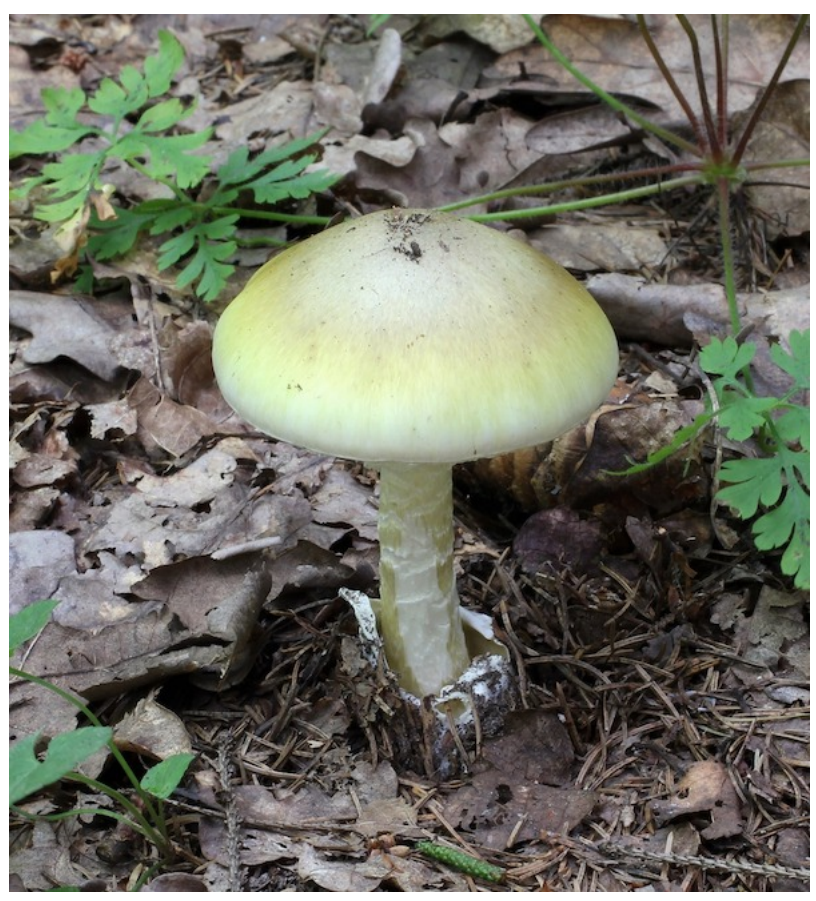

Fig. 1.3.2 Death cap - do not eat!

Classification can get much more complicated than just binary or multiclass classification. For instance, there are some variants of classification addressing hierarchically structured classes. In such cases not all errors are equal—if we must err, we might prefer to misclassify to a related class rather than a distant class. Usually, this is referred to as *hierarchical classification*. For inspiration, you might think of Linnaeus[20], who organized fauna in a hierarchy.

In the case of animal classification, it might not be so bad to mistake a poodle for a schnauzer, but our model would pay a huge penalty if it confused a poodle with a dinosaur. Which hierarchy is relevant might depend on how you plan to use the model. For example, rattlesnakes and garter snakes might be close on the phylogenetic tree, but mistaking a rattler for a garter could have fatal consequences.

### Tagging

Some classification problems fit neatly into the binary or multiclass classification setups. For example, we could train a normal binary classifier to distinguish cats from dogs. Given the current state of computer vision, we can do this easily, with off-the-shelf tools. Nonetheless, no matter how accurate our model gets, we might find ourselves in trouble when the classifier encounters an image of the *Town Musicians of Bremen*, a popular German fairy tale featuring four animals (Fig. 1.3.3).

As you can see, the photo features a cat, a rooster, a dog, and a donkey, with some trees in the background. If we anticipate encountering such images, multiclass classification might not be the right problem formulation. Instead, we might want to give the model the option of saying the image depicts a cat, a dog, a donkey, *and* a rooster.

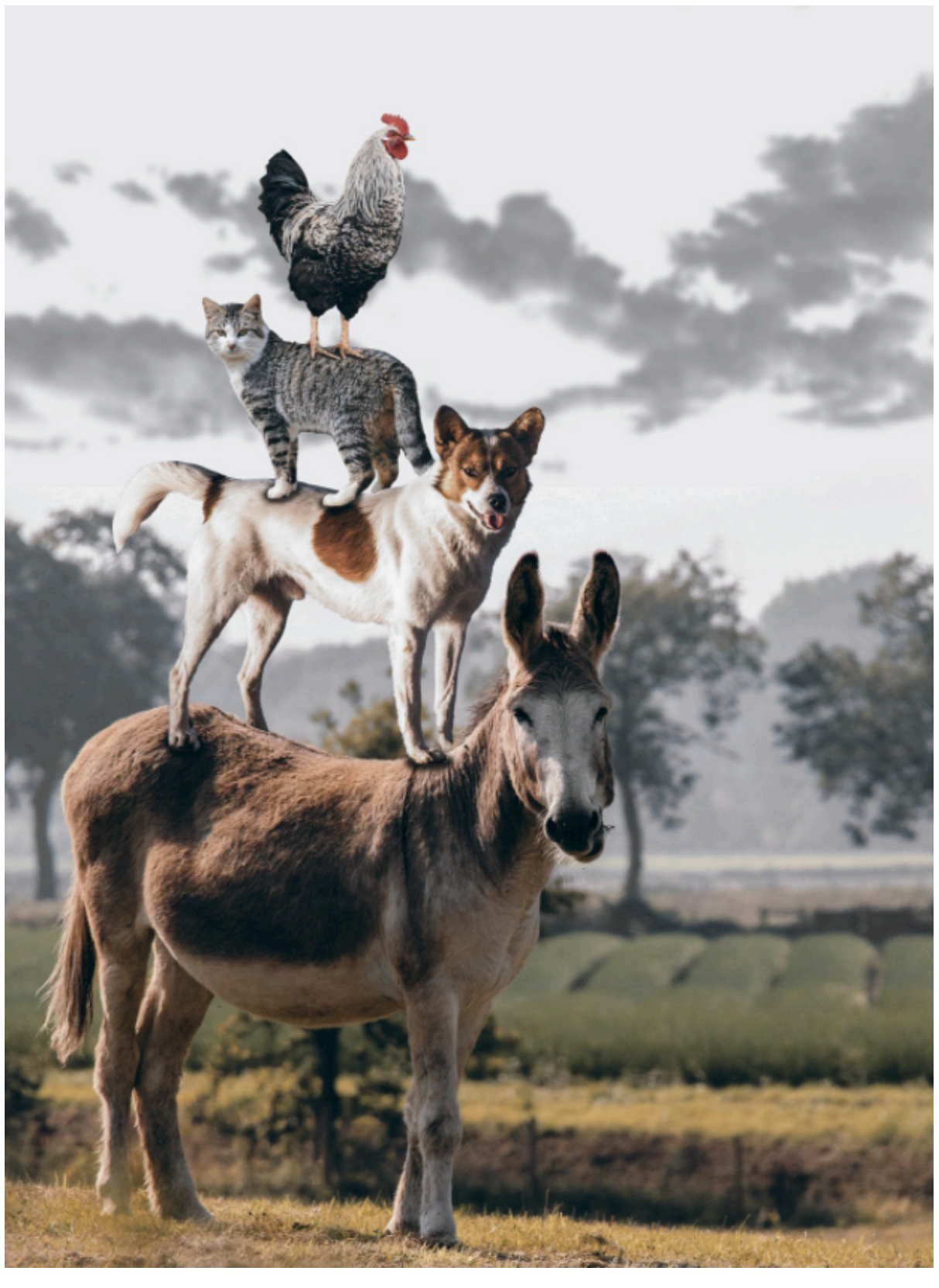

Fig. 1.3.3 A donkey, a dog, a cat, and a rooster.

The problem of learning to predict classes that are not mutually exclusive is called *multi-label classification*. Auto-tagging problems are typically best described in terms of multi-label classification. Think of the tags people might apply to posts on a technical blog, e.g., “machine learning”, “technology”, “gadgets”, “programming languages”, “Linux”, “cloud computing”, “AWS”. A typical article might have 5–10 tags applied. Typically, tags will exhibit some correlation structure. Posts about “cloud computing” are likely to mention “AWS” and posts about “machine learning” are likely to mention “GPUs”.

Sometimes such tagging problems draw on enormous label sets. The National Library of Medicine employs many professional annotators who associate each article to be indexed in PubMed with a set of tags drawn from the Medical Subject Headings (MeSH) ontology, a collection of roughly 28,000 tags. Correctly tagging articles is important because it allows researchers to conduct exhaustive reviews of the literature. This is a time-consuming process and typically there is a one-year lag between archiving and tagging. Machine learning can provide provisional tags until each article has a proper manual review. Indeed, for several years, the BioASQ organization has hosted competitions[21] for this task.

### Search

In the field of information retrieval, we often impose ranks on sets of items. Take web search for example. The goal is less to determine *whether* a particular page is relevant for a query, but rather which, among a set of relevant results, should be shown most prominently to a particular user. One way of doing this might be to first assign a score to every element in the set and then to retrieve the top-rated elements. PageRank [22], the original secret sauce behind the Google search engine, was an early example of such a scoring system. Weirdly, the scoring provided by PageRank did not depend on the actual query. Instead, they relied on a simple relevance filter to identify the set of relevant candidates and then used PageRank to prioritize the more authoritative pages. Nowadays, search engines use machine learning and behavioral models to obtain query-dependent relevance scores. There are entire academic conferences devoted to this subject.

22

### Recommender Systems

Recommender systems are another problem setting that is related to search and ranking. The problems are similar insofar as the goal is to display a set of items relevant to the user. The main difference is the emphasis on *personalization* to specific users in the context of recommender systems. For instance, for movie recommendations, the results page for a science fiction fan and the results page for a connoisseur of Peter Sellers comedies might differ significantly. Similar problems pop up in other recommendation settings, e.g., for retail products, music, and news recommendation.

In some cases, customers provide explicit feedback, communicating how much they liked a particular product (e.g., the product ratings and reviews on Amazon, IMDb, or Goodreads). In other cases, they provide implicit feedback, e.g., by skipping titles on a playlist, which might indicate dissatisfaction or maybe just indicate that the song was inappropriate in context. In the simplest formulations, these systems are trained to estimate some score, such as an expected star rating or the probability that a given user will purchase a particular item.

Given such a model, for any given user, we could retrieve the set of objects with the largest scores, which could then be recommended to the user. Production systems are considerably more advanced and take detailed user activity and item characteristics into account when computing such scores. Fig. 1.3.4 displays the deep learning books recommended by Amazon based on personalization algorithms tuned to capture Aston's preferences.

Despite their tremendous economic value, recommender systems naively built on top of predictive models suffer some serious conceptual flaws. To start, we only observe *censored feedback*: users preferentially rate movies that they feel strongly about. For example, on a five-point scale, you might notice that items receive many one- and five-star ratings but that there are conspicuously few three-star ratings. Moreover, current purchase habits are often a result of the recommendation algorithm currently in place, but learning algorithms do not always take this detail into account. Thus it is possible for feedback loops to form where a recommender system preferentially pushes an item that is then taken to be better (due to greater pur-chases) and in turn is recommended even more frequently. Many of these problems—about how to deal with censoring, incentives, and feedback loops—are important open research questions.

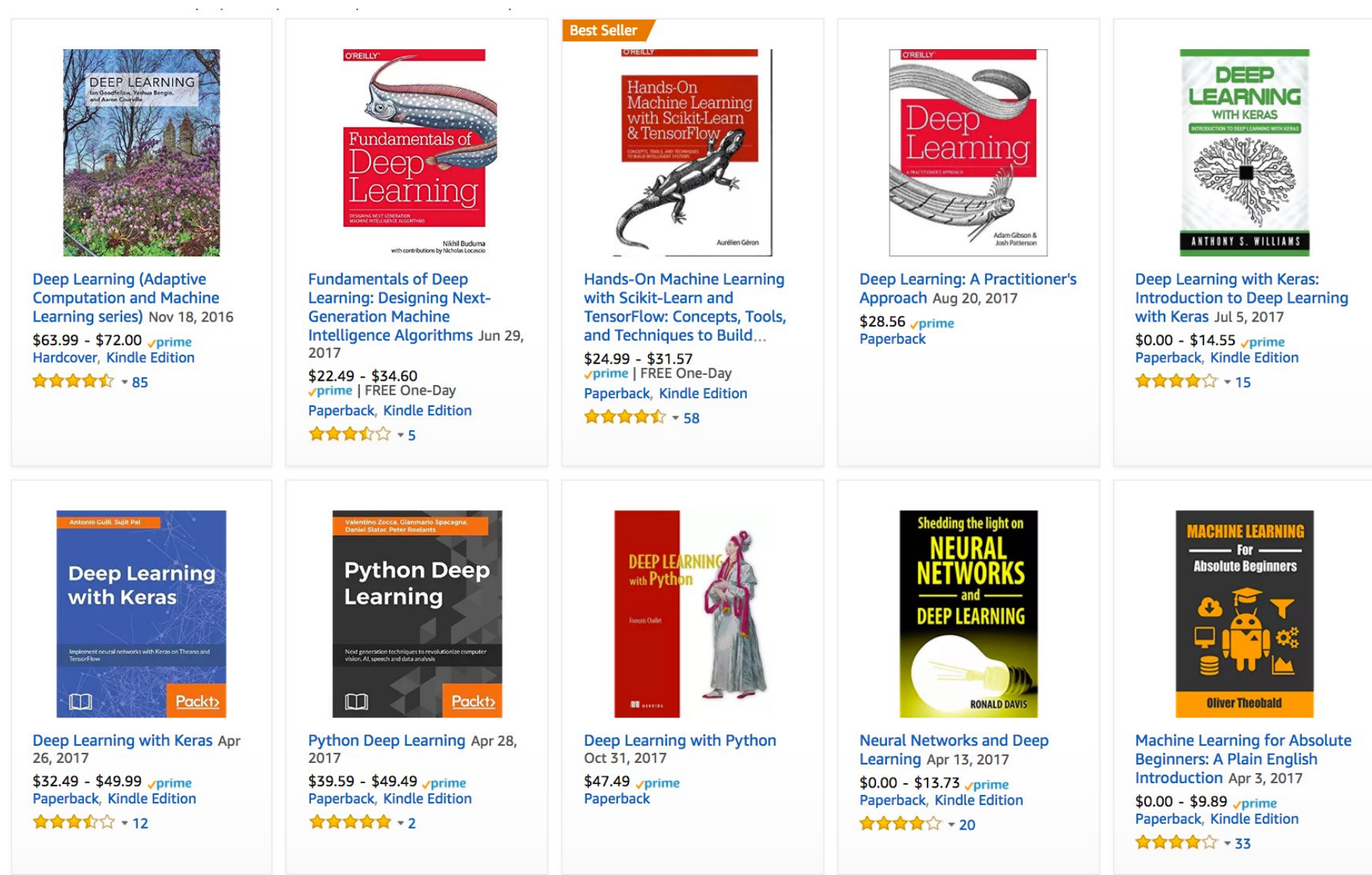

Fig. 1.3.4 Deep learning books recommended by Amazon.

### Sequence Learning

So far, we have looked at problems where we have some fixed number of inputs and produce a fixed number of outputs. For example, we considered predicting house prices given a fixed set of features: square footage, number of bedrooms, number of bathrooms, and the transit time to downtown. We also discussed mapping from an image (of fixed dimension) to the predicted probabilities that it belongs to each among a fixed number of classes and predicting star ratings associated with purchases based on the user ID and product ID alone. In these cases, once our model is trained, after each test example is fed into our model, it is immediately forgotten. We assumed that successive observations were independent and thus there was no need to hold on to this context.

But how should we deal with video snippets? In this case, each snippet might consist of a different number of frames. And our guess of what is going on in each frame might be much stronger if we take into account the previous or succeeding frames. The same goes for language. For example, one popular deep learning problem is machine translation: the task of ingesting sentences in some source language and predicting their translations in another language.

Such problems also occur in medicine. We might want a model to monitor patients in the intensive care unit and to fire off alerts whenever their risk of dying in the next 24 hours exceeds some threshold. Here, we would not throw away everything that we know about the patient history every hour, because we might not want to make predictions based only on the most recent measurements.

Questions like these are among the most exciting applications of machine learning and they are instances of *sequence learning*. They require a model either to ingest sequences of inputs or to emit sequences of outputs (or both). Specifically, *sequence-to-sequence learning* considers problems where both inputs and outputs consist of variable-length sequences. Examples include machine translation and speech-to-text transcription. While it is impossible to consider all types of sequence transformations, the following special cases are worth mentioning.

**Tagging and Parsing**. This involves annotating a text sequence with attributes. Here, the inputs and outputs are *aligned*, i.e., they are of the same number and occur in a corresponding order. For instance, in *part-of-speech (PoS) tagging*, we annotate every word in a sentence with the corresponding part of speech, i.e., "noun" or "direct object". Alternatively, we might want to know which groups of contiguous words refer to named entities, like *people*, *places*, or *organizations*. In the cartoonishly simple example below, we might just want to indicate whether or not any word in the sentence is part of a named entity (tagged as "Ent").

```
Tom has dinner in Washington with Sally
Ent  -    -    -     Ent      -    Ent
```

**Automatic Speech Recognition**. With speech recognition, the input sequence is an audio recording of a speaker (Fig. 1.3.5), and the output is a transcript of what the speaker said. The challenge is that there are many more audio frames (sound is typically sampled at 8kHz or 16kHz) than text, i.e., there is no 1:1 correspondence between audio and text, since thousands of samples may correspond to a single spoken word. These are sequence-to-sequence learning problems, where the output is much shorter than the input. While humans are remarkably good at recognizing speech, even from low-quality audio, getting computers to perform the same feat is a formidable challenge.

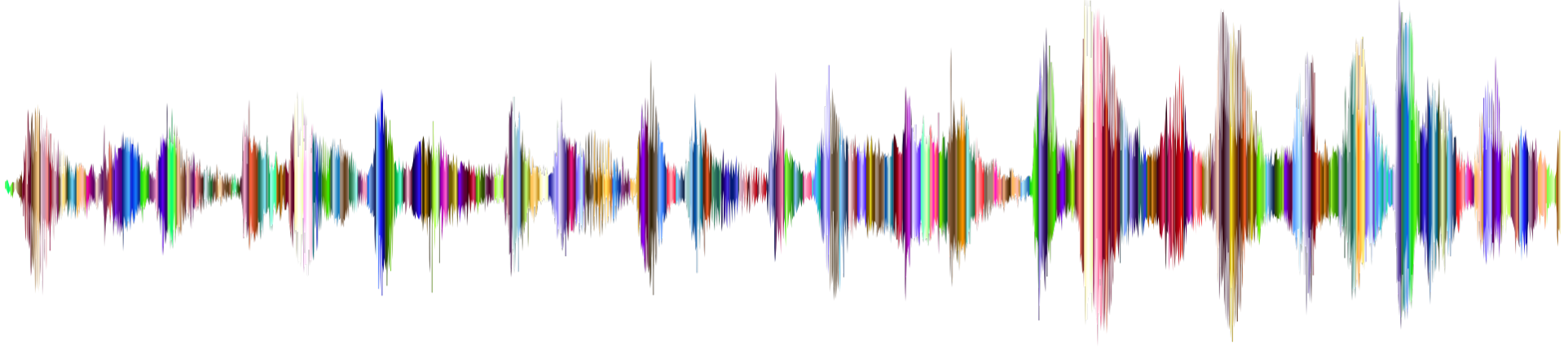

Fig. 1.3.5 `-D-e-e-p- L-ea-r-ni-ng-` in an audio recording.

**Text to Speech**. This is the inverse of automatic speech recognition. Here, the input is text and the output is an audio file. In this case, the output is much longer than the input.

**Machine Translation**. Unlike the case of speech recognition, where corresponding inputs and outputs occur in the same order, in machine translation, unaligned data poses a new chal-lenge. Here the input and output sequences can have different lengths, and the corresponding regions of the respective sequences may appear in a different order. Consider the following illustrative example of the peculiar tendency of Germans to place the verbs at the end of sentences:

```
German:           Haben Sie sich schon dieses grossartige Lehrwerk angeschaut?
English:          Have you already looked at this excellent textbook?
Wrong alignment:  Have you yourself already this excellent textbook looked at?
```

Many related problems pop up in other learning tasks. For instance, determining the order in which a user reads a webpage is a two-dimensional layout analysis problem. Dialogue problems exhibit all kinds of additional complications, where determining what to say next requires taking into account real-world knowledge and the prior state of the conversation across long temporal distances. Such topics are active areas of research.

### 1.3.2 Unsupervised and Self-Supervised Learning

The previous examples focused on supervised learning, where we feed the model a giant dataset containing both the features and corresponding label values. You could think of the supervised learner as having an extremely specialized job and an extremely dictatorial boss. The boss stands over the learner's shoulder and tells them exactly what to do in every situation until they learn to map from situations to actions. Working for such a boss sounds pretty lame. On the other hand, pleasing such a boss is pretty easy. You just recognize the pattern as quickly as possible and imitate the boss's actions.

Considering the opposite situation, it could be frustrating to work for a boss who has no idea what they want you to do. However, if you plan to be a data scientist, you had better get used to it. The boss might just hand you a giant dump of data and tell you to *do some data science with it!* This sounds vague because it is vague. We call this class of problems *unsupervised learning*, and the type and number of questions we can ask is limited only by our creativity. We will address unsupervised learning techniques in later chapters. To whet your appetite for now, we describe a few of the following questions you might ask.

- Can we find a small number of prototypes that accurately summarize the data? Given a set of photos, can we group them into landscape photos, pictures of dogs, babies, cats, and mountain peaks? Likewise, given a collection of users' browsing activities, can we group them into users with similar behavior? This problem is typically known as *clustering*.
- Can we find a small number of parameters that accurately capture the relevant properties of the data? The trajectories of a ball are well described by velocity, diameter, and mass of the ball. Tailors have developed a small number of parameters that describe human body shape fairly accurately for the purpose of fitting clothes. These problems are referred to as *subspace estimation*. If the dependence is linear, it is called *principal component analysis*.
- Is there a representation of (arbitrarily structured) objects in Euclidean space such that

symbolic properties can be well matched? This can be used to describe entities and their relations, such as "Rome" − "Italy" + "France" = "Paris".

- Is there a description of the root causes of much of the data that we observe? For instance, if we have demographic data about house prices, pollution, crime, location, education, and salaries, can we discover how they are related simply based on empirical data? The fields concerned with *causality* and *probabilistic graphical models* tackle such questions.
- Another important and exciting recent development in unsupervised learning is the advent of *deep generative models*. These models estimate the density of the data, either explicitly or *implicitly*. Once trained, we can use a generative model either to score examples according to how likely they are, or to sample synthetic examples from the learned distribution. Early deep learning breakthroughs in generative modeling came with the invention of *variational autoencoders* (Kingma and Welling, 2014, Rezende *et al.*, 2014) and continued with the development of *generative adversarial networks* (Goodfellow *et al.*, 2014). More recent advances include normalizing flows (Dinh *et al.*, 2014, Dinh *et al.*, 2017) and diffusion models (Ho *et al.*, 2020, Sohl-Dickstein *et al.*, 2015, Song and Ermon, 2019, Song *et al.*, 2021).

A further development in unsupervised learning has been the rise of *self-supervised learning*, techniques that leverage some aspect of the unlabeled data to provide supervision. For text, we can train models to "fill in the blanks" by predicting randomly masked words using their surrounding words (contexts) in big corpora without any labeling effort (Devlin *et al.*, 2018)! For images, we may train models to tell the relative position between two cropped regions of the same image (Doersch *et al.*, 2015), to predict an occluded part of an image based on the remaining portions of the image, or to predict whether two examples are perturbed versions of the same underlying image. Self-supervised models often learn representations that are subsequently leveraged by fine-tuning the resulting models on some downstream task of interest.

### 1.3.3 Interacting with an Environment

So far, we have not discussed where data actually comes from, or what actually happens when a machine learning model generates an output. That is because supervised learning and unsupervised learning do not address these issues in a very sophisticated way. In each case, we grab a big pile of data upfront, then set our pattern recognition machines in motion without ever interacting with the environment again. Because all the learning takes place after the algorithm is disconnected from the environment, this is sometimes called *offline learning*. For example, supervised learning assumes the simple interaction pattern depicted in Fig. 1.3.6.

This simplicity of offline learning has its charms. The upside is that we can worry about pattern recognition in isolation, with no concern about complications arising from interactions with a dynamic environment. But this problem formulation is limiting. If you grew up reading Asimov's Robot novels, then you probably picture artificially intelligent agents capable not only of making predictions, but also of taking actions in the world. We want to think about intelligent *agents*, not just predictive models. This means that we need to think about choosing *actions*, not just making predictions. In contrast to mere predictions, actions actually impact the environment. If we want to train an intelligent agent, we must account for the way its actions might impact the future observations of the agent, and so offline learning is inappropriate.

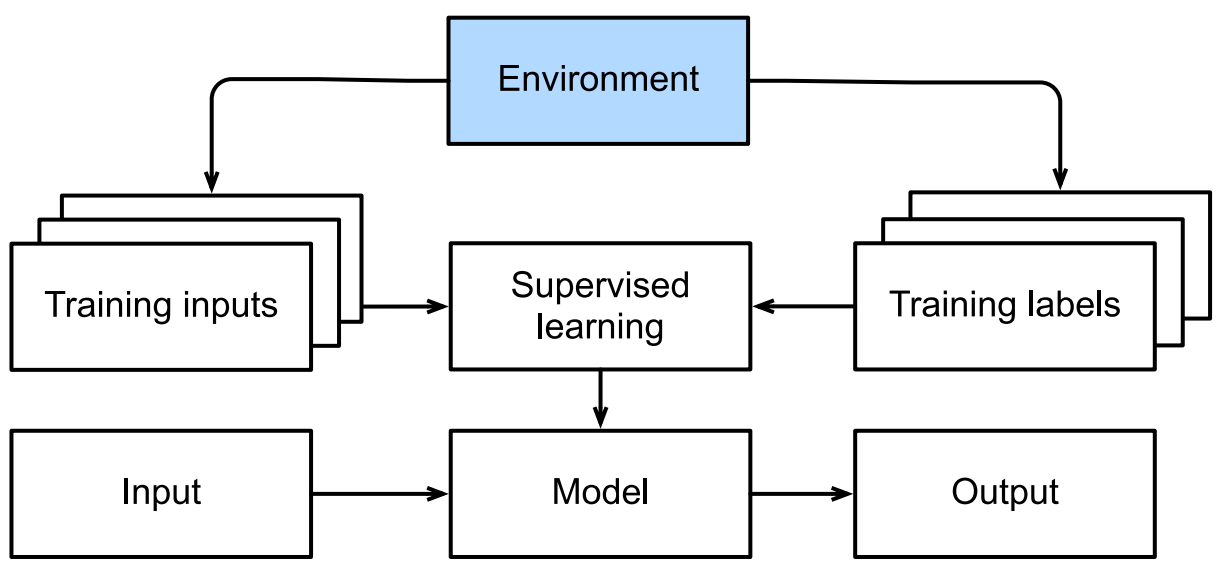

Fig. 1.3.6 Collecting data for supervised learning from an environment.

Considering the interaction with an environment opens a whole set of new modeling questions. The following are just a few examples.

- Does the environment remember what we did previously?
- Does the environment want to help us, e.g., a user reading text into a speech recognizer?
- Does the environment want to beat us, e.g., spammers adapting their emails to evade spam filters?
- Does the environment have shifting dynamics? For example, would future data always resemble the past or would the patterns change over time, either naturally or in response to our automated tools?

These questions raise the problem of *distribution shift*, where training and test data are different. An example of this, that many of us may have met, is when taking exams written by a lecturer, while the homework was composed by their teaching assistants. Next, we briefly describe reinforcement learning, a rich framework for posing learning problems in which an agent interacts with an environment.

### 1.3.4 Reinforcement Learning

If you are interested in using machine learning to develop an agent that interacts with an environment and takes actions, then you are probably going to wind up focusing on *reinforcement learning*. This might include applications to robotics, to dialogue systems, and even to developing artificial intelligence (AI) for video games. *Deep reinforcement learning*, which applies deep learning to reinforcement learning problems, has surged in popularity. The breakthrough deep Q-network, that beat humans at Atari games using only the visual input (Mnih *et al.*, 2015), and the AlphaGo program, which dethroned the world champion at the board game Go (Silver *et al.*, 2016), are two prominent examples.

Reinforcement learning gives a very general statement of a problem in which an agent interacts with an environment over a series of time steps. At each time step, the agent receives some *observation* from the environment and must choose an *action* that is subsequently transmitted back to the environment via some mechanism (sometimes called an *actuator*), when, after each loop, the agent receives a reward from the environment. This process is illustrated in Fig. 1.3.7. The agent then receives a subsequent observation, and chooses a subsequent action, and so on. The behavior of a reinforcement learning agent is governed by a *policy*. In brief, a *policy* is just a function that maps from observations of the environment to actions. The goal of reinforcement learning is to produce good policies.

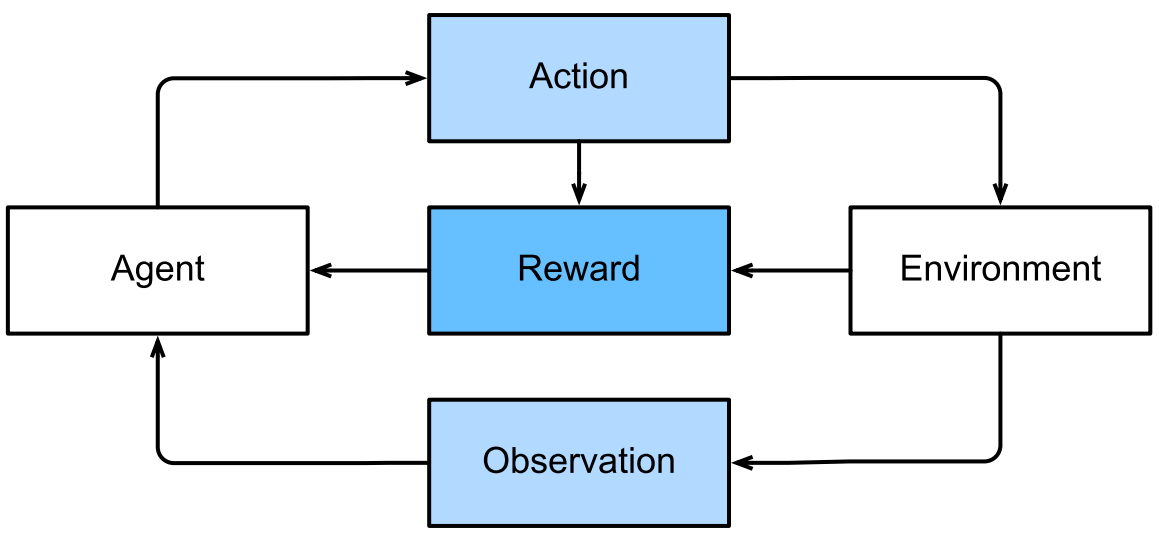

Fig. 1.3.7 The interaction between reinforcement learning and an environment.

It is hard to overstate the generality of the reinforcement learning framework. For example, supervised learning can be recast as reinforcement learning. Say we had a classification problem. We could create a reinforcement learning agent with one action corresponding to each class. We could then create an environment which gave a reward that was exactly equal to the loss function from the original supervised learning problem.

Further, reinforcement learning can also address many problems that supervised learning cannot. For example, in supervised learning, we always expect that the training input comes associated with the correct label. But in reinforcement learning, we do not assume that, for each observation the environment tells us the optimal action. In general, we just get some reward. Moreover, the environment may not even tell us which actions led to the reward.

Consider the game of chess. The only real reward signal comes at the end of the game when we either win, earning a reward of, say, $1$, or when we lose, receiving a reward of, say, $-1$. So reinforcement learners must deal with the *credit assignment* problem: determining which actions to credit or blame for an outcome. The same goes for an employee who gets a promotion on October 11. That promotion likely reflects a number of well-chosen actions over the previous year. Getting promoted in the future requires figuring out which actions along the way led to the earlier promotions.

Reinforcement learners may also have to deal with the problem of partial observability. That is, the current observation might not tell you everything about your current state. Say your cleaning robot found itself trapped in one of many identical closets in your house. Rescuing the robot involves inferring its precise location which might require considering earlier observations prior to it entering the closet.

Finally, at any given point, reinforcement learners might know of one good policy, but there might be many other better policies that the agent has never tried. The reinforcement learner must constantly choose whether to *exploit* the best (currently) known strategy as a policy, or to *explore* the space of strategies, potentially giving up some short-term reward in exchange for knowledge.

The general reinforcement learning problem has a very general setting. Actions affect subsequent observations. Rewards are only observed when they correspond to the chosen actions. The environment may be either fully or partially observed. Accounting for all this complexity at once may be asking too much. Moreover, not every practical problem exhibits all this complexity. As a result, researchers have studied a number of special cases of reinforcement learning problems.

When the environment is fully observed, we call the reinforcement learning problem a *Markov decision process*. When the state does not depend on the previous actions, we call it a *contextual bandit problem*. When there is no state, just a set of available actions with initially unknown rewards, we have the classic *multi-armed bandit problem*.

## 1.4 Roots

We have just reviewed a small subset of problems that machine learning can address. For a diverse set of machine learning problems, deep learning provides powerful tools for their solution. Although many deep learning methods are recent inventions, the core ideas behind learning from data have been studied for centuries. In fact, humans have held the desire to analyze data and to predict future outcomes for ages, and it is this desire that is at the root of much of natural science and mathematics. Two examples are the Bernoulli distribution, named after Jacob Bernoulli (1655–1705)[23], and the Gaussian distribution discovered by Carl Friedrich Gauss (1777–1855)[24]. Gauss invented, for instance, the least mean squares algorithm, which is still used today for a multitude of problems from insurance calculations to medical diagnostics. Such tools enhanced the experimental approach in the natural sciences—for instance, Ohm's law relating current and voltage in a resistor is perfectly described by a linear model.

23 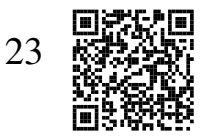

24 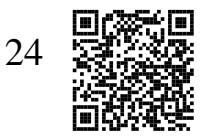

Even in the middle ages, mathematicians had a keen intuition of estimates. For instance, the geometry book of Jacob Köbel (1460–1533)[25] illustrates averaging the length of 16 adult men's feet to estimate the typical foot length in the population (Fig. 1.4.1).

As a group of individuals exited a church, 16 adult men were asked to line up in a row and have their feet measured. The sum of these measurements was then divided by 16 to obtain an estimate for what now is called one foot. This "algorithm" was later improved to deal with misshapen feet; The two men with the shortest and longest feet were sent away, averaging only over the remainder. This is among the earliest examples of a trimmed mean estimate.

Statistics really took off with the availability and collection of data. One of its pioneers,

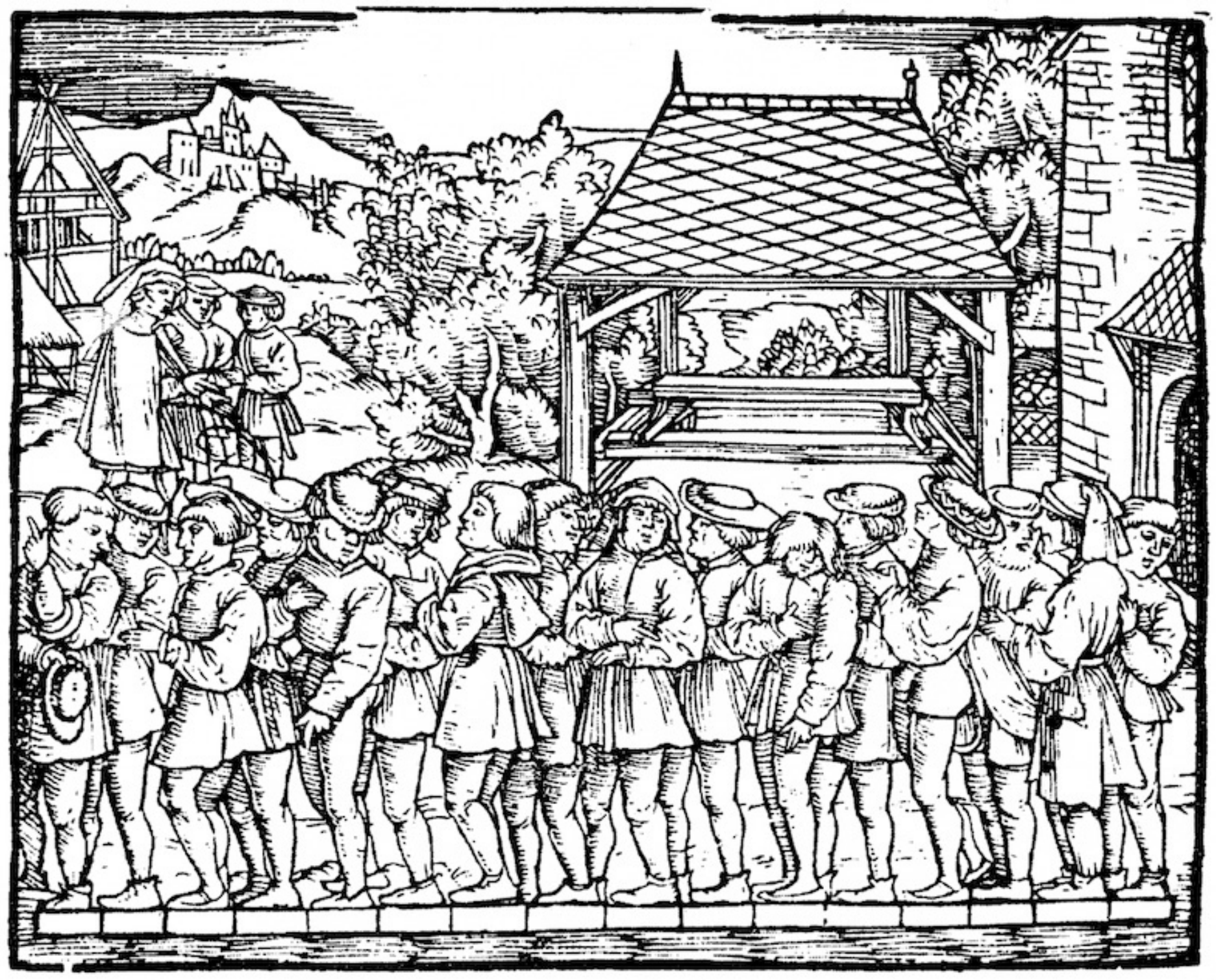

Fig. 1.4.1 Estimating the length of a foot.

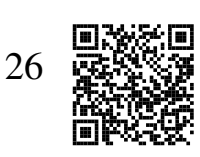

Ronald Fisher (1890–1962)[26], contributed significantly to its theory and also its applications in genetics. Many of his algorithms (such as linear discriminant analysis) and concepts (such as the Fisher information matrix) still hold a prominent place in the foundations of modern statistics. Even his data resources had a lasting impact. The Iris dataset that Fisher released in 1936 is still sometimes used to demonstrate machine learning algorithms. Fisher was also a proponent of eugenics, which should remind us that the morally dubious use of data science has as long and enduring a history as its productive use in industry and the natural sciences.

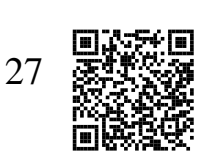

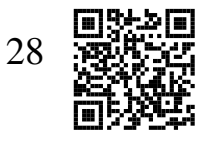

Other influences for machine learning came from the information theory of Claude Shannon (1916–2001)[27] and the theory of computation proposed by Alan Turing (1912–1954)[28]. Turing posed the question "can machines think?" in his famous paper *Computing Machinery and Intelligence* (Turing, 1950). Describing what is now known as the Turing test, he proposed that a machine can be considered *intelligent* if it is difficult for a human evaluator to distinguish between the replies from a machine and those of a human, based purely on textual interactions.

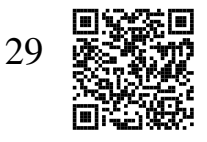

Further influences came from neuroscience and psychology. After all, humans clearly exhibit intelligent behavior. Many scholars have asked whether one could explain and possibly reverse engineer this capacity. One of the first biologically inspired algorithms was formulated by Donald Hebb (1904–1985)[29]. In his groundbreaking book *The Organization of Behavior* (Hebb, 1949), he posited that neurons learn by positive reinforcement. This became known as the Hebbian learning rule. These ideas inspired later work, such as Rosenblatt's perceptron learning algorithm, and laid the foundations of many stochastic gradient descent algorithms that underpin deep learning today: reinforce desirable behavior and diminish undesirable behavior to obtain good settings of the parameters in a neural network.

Biological inspiration is what gave *neural networks* their name. For over a century (dating back to the models of Alexander Bain, 1873, and James Sherrington, 1890), researchers have tried to assemble computational circuits that resemble networks of interacting neurons. Over time, the interpretation of biology has become less literal, but the name stuck. At its heart lie a few key principles that can be found in most networks today:

- The alternation of linear and nonlinear processing units, often referred to as *layers*.
- The use of the chain rule (also known as *backpropagation*) for adjusting parameters in the entire network at once.

After initial rapid progress, research in neural networks languished from around 1995 until 2005. This was mainly due to two reasons. First, training a network is computationally very expensive. While random-access memory was plentiful at the end of the past century, computational power was scarce. Second, datasets were relatively small. In fact, Fisher's Iris dataset from 1936 was still a popular tool for testing the efficacy of algorithms. The MNIST dataset with its 60,000 handwritten digits was considered huge.

Given the scarcity of data and computation, strong statistical tools such as kernel methods, decision trees, and graphical models proved empirically superior in many applications. Moreover, unlike neural networks, they did not require weeks to train and provided predictable results with strong theoretical guarantees.

## 1.5 The Road to Deep Learning

Much of this changed with the availability of massive amounts of data, thanks to the World Wide Web, the advent of companies serving hundreds of millions of users online, a dissemination of low-cost, high-quality sensors, inexpensive data storage (Kryder's law), and cheap computation (Moore's law). In particular, the landscape of computation in deep learning was revolutionized by advances in GPUs that were originally engineered for computer gaming. Suddenly algorithms and models that seemed computationally infeasible were within reach. This is best illustrated in Table 1.5.1.

Table 1.5.1: Dataset vs. computer memory and computational power

| Decade | Dataset | Memory | Floating point calculations per second |
|---|---|---|---|
| 1970 | 100 (Iris) | 1 KB | 100 KF (Intel 8080) |
| 1980 | 1 K (house prices in Boston) | 100 KB | 1 MF (Intel 80186) |
| 1990 | 10 K (optical character recognition) | 10 MB | 10 MF (Intel 80486) |
| 2000 | 10 M (web pages) | 100 MB | 1 GF (Intel Core) |
| 2010 | 10 G (advertising) | 1 GB | 1 TF (NVIDIA C2050) |
| 2020 | 1 T (social network) | 100 GB | 1 PF (NVIDIA DGX-2) |

Note that random-access memory has not kept pace with the growth in data. At the same time, increases in computational power have outpaced the growth in datasets. This means that statistical models need to become more memory efficient, and so they are free to spend more computer cycles optimizing parameters, thanks to the increased compute budget. Consequently, the sweet spot in machine learning and statistics moved from (generalized) linear models and kernel methods to deep neural networks. This is also one of the reasons why many of the mainstays of deep learning, such as multilayer perceptrons (McCulloch and Pitts, 1943), convolutional neural networks (LeCun *et al.*, 1998), long short-term memory (Hochreiter and Schmidhuber, 1997), and Q-Learning (Watkins and Dayan, 1992), were essentially "rediscovered" in the past decade, after lying comparatively dormant for considerable time.

The recent progress in statistical models, applications, and algorithms has sometimes been likened to the Cambrian explosion: a moment of rapid progress in the evolution of species. Indeed, the state of the art is not just a mere consequence of available resources applied to decades-old algorithms. Note that the list of ideas below barely scratches the surface of what has helped researchers achieve tremendous progress over the past decade.

- Novel methods for capacity control, such as *dropout* (Srivastava *et al.*, 2014), have helped to mitigate overfitting. Here, noise is injected (Bishop, 1995) throughout the neural network during training.
- *Attention mechanisms* solved a second problem that had plagued statistics for over a century: how to increase the memory and complexity of a system without increasing the number of learnable parameters. Researchers found an elegant solution by using what can only be viewed as a *learnable pointer structure* (Bahdanau *et al.*, 2014). Rather than having to remember an entire text sequence, e.g., for machine translation in a fixed-dimensional representation, all that needed to be stored was a pointer to the intermediate state of the translation process. This allowed for significantly increased accuracy for long sequences, since the model no longer needed to remember the entire sequence before commencing the generation of a new one.
- Built solely on attention mechanisms, the *Transformer* architecture (Vaswani *et al.*, 2017)

has demonstrated superior *scaling* behavior: it performs better with an increase in dataset size, model size, and amount of training compute (Kaplan *et al.*, 2020). This architecture has demonstrated compelling success in a wide range of areas, such as natural language processing (Brown *et al.*, 2020, Devlin *et al.*, 2018), computer vision (Dosovitskiy *et al.*, 2021, Liu *et al.*, 2021), speech recognition (Gulati *et al.*, 2020), reinforcement learning (Chen *et al.*, 2021), and graph neural networks (Dwivedi and Bresson, 2020). For example, a single Transformer pretrained on modalities as diverse as text, images, joint torques, and button presses can play Atari, caption images, chat, and control a robot (Reed *et al.*, 2022).

- Modeling probabilities of text sequences, *language models* can predict text given other text. Scaling up the data, model, and compute has unlocked a growing number of capabilities of language models to perform desired tasks via human-like text generation based on input text (Anil *et al.*, 2023, Brown *et al.*, 2020, Chowdhery *et al.*, 2022, Hoffmann *et al.*, 2022, OpenAI, 2023, Rae *et al.*, 2021, Touvron *et al.*, 2023a, Touvron *et al.*, 2023b). For instance, aligning language models with human intent (Ouyang *et al.*, 2022), OpenAI's ChatGPT[30] allows users to interact with it in a conversational way to solve problems, such as code debugging and creative writing.

30

- Multi-stage designs, e.g., via the memory networks (Sukhbaatar *et al.*, 2015) and the neural programmer-interpreter (Reed and De Freitas, 2015) permitted statistical modelers to describe iterative approaches to reasoning. These tools allow for an internal state of the deep neural network to be modified repeatedly, thus carrying out subsequent steps in a chain of reasoning, just as a processor can modify memory for a computation.

- A key development in *deep generative modeling* was the invention of *generative adversarial networks* (Goodfellow *et al.*, 2014). Traditionally, statistical methods for density estimation and generative models focused on finding proper probability distributions and (often approximate) algorithms for sampling from them. As a result, these algorithms were largely limited by the lack of flexibility inherent in the statistical models. The crucial innovation in generative adversarial networks was to replace the sampler by an arbitrary algorithm with differentiable parameters. These are then adjusted in such a way that the discriminator (effectively a two-sample test) cannot distinguish fake from real data. Through the ability to use arbitrary algorithms to generate data, density estimation was opened up to a wide variety of techniques. Examples of galloping zebras (Zhu *et al.*, 2017) and of fake celebrity faces (Karras *et al.*, 2017) are each testimony to this progress. Even amateur doodlers can produce photorealistic images just based on sketches describing the layout of a scene (Park *et al.*, 2019).

- Furthermore, while the diffusion process gradually adds random noise to data samples, *diffusion models* (Ho *et al.*, 2020, Sohl-Dickstein *et al.*, 2015) learn the denoising process to gradually construct data samples from random noise, reversing the diffusion process. They have started to replace generative adversarial networks in more recent deep generative models, such as in DALL-E 2 (Ramesh *et al.*, 2022) and Imagen (Saharia *et al.*, 2022) for creative art and image generation based on text descriptions.

- In many cases, a single GPU is insufficient for processing the large amounts of data avail-

able for training. Over the past decade the ability to build parallel and distributed training algorithms has improved significantly. One of the key challenges in designing scalable algorithms is that the workhorse of deep learning optimization, stochastic gradient descent, relies on relatively small minibatches of data to be processed. At the same time, small batches limit the efficiency of GPUs. Hence, training on 1,024 GPUs with a minibatch size of, say, 32 images per batch amounts to an aggregate minibatch of about 32,000 images. Work, first by Li (2017) and subsequently by You *et al.* (2017) and Jia *et al.* (2018) pushed the size up to 64,000 observations, reducing training time for the ResNet-50 model on the ImageNet dataset to less than 7 minutes. By comparison, training times were initially of the order of days.

31 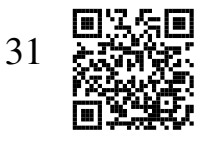

32 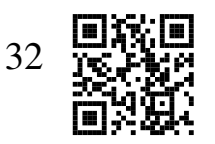

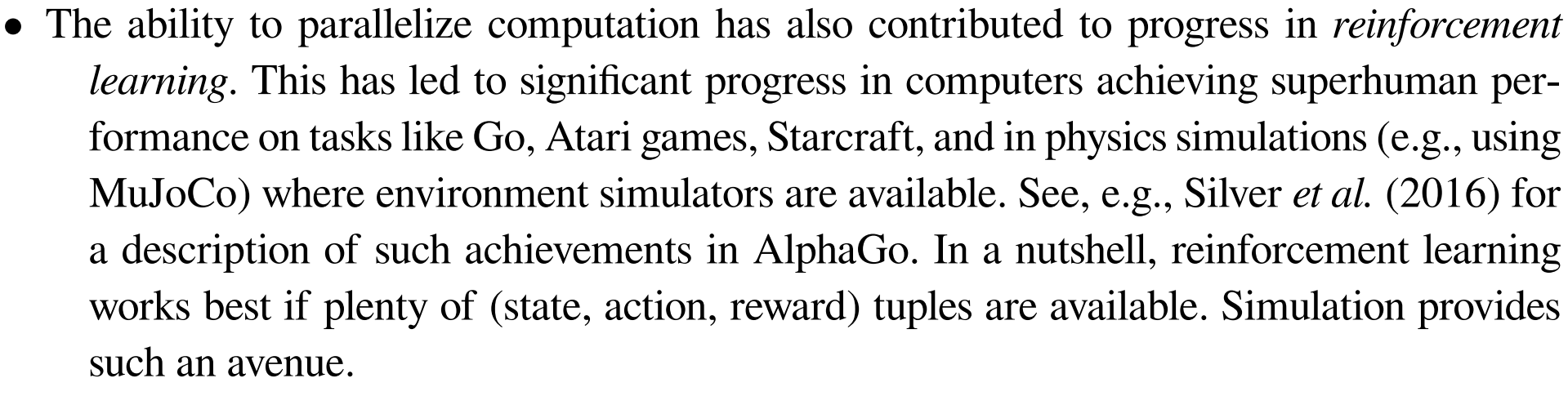

- The ability to parallelize computation has also contributed to progress in *reinforcement learning*. This has led to significant progress in computers achieving superhuman performance on tasks like Go, Atari games, Starcraft, and in physics simulations (e.g., using MuJoCo) where environment simulators are available. See, e.g., Silver *et al.* (2016) for a description of such achievements in AlphaGo. In a nutshell, reinforcement learning works best if plenty of (state, action, reward) tuples are available. Simulation provides such an avenue.

33

34

35

- Deep learning frameworks have played a crucial role in disseminating ideas. The first generation of open-source frameworks for neural network modeling consisted of Caffe[31], Torch[32], and Theano[33]. Many seminal papers were written using these tools. These have now been superseded by TensorFlow[34] (often used via its high-level API Keras[35]), CNTK[36], Caffe 2[37], and Apache MXNet[38]. The third generation of frameworks consists of so-called *imperative* tools for deep learning, a trend that was arguably ignited by Chainer[39], which used a syntax similar to Python NumPy to describe models. This idea was adopted by both PyTorch[40], the Gluon API[41] of MXNet, and JAX[42].

36

37 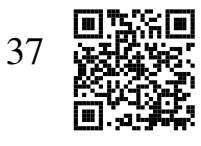

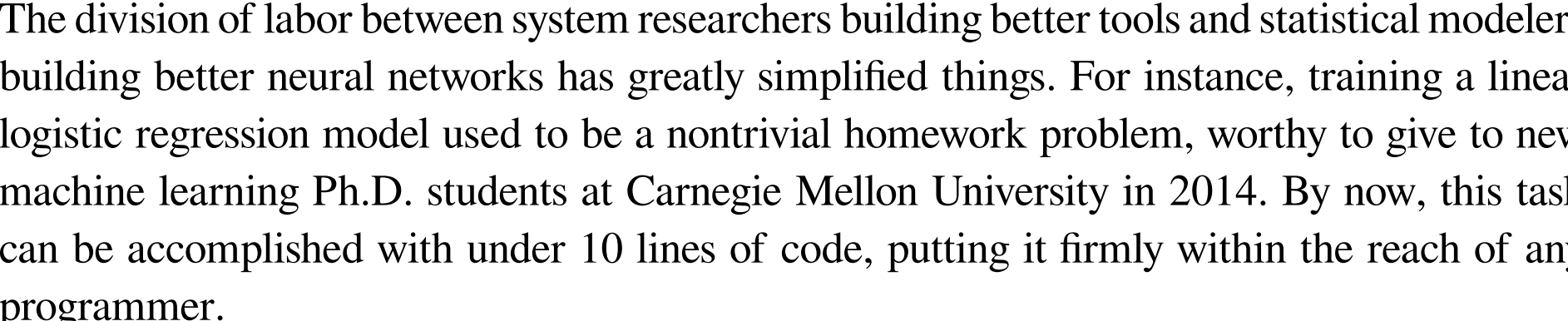

The division of labor between system researchers building better tools and statistical modelers building better neural networks has greatly simplified things. For instance, training a linear logistic regression model used to be a nontrivial homework problem, worthy to give to new machine learning Ph.D. students at Carnegie Mellon University in 2014. By now, this task can be accomplished with under 10 lines of code, putting it firmly within the reach of any programmer.

39

40

## 1.6 Success Stories

41 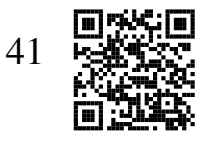

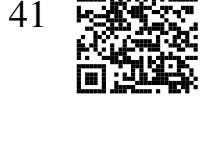

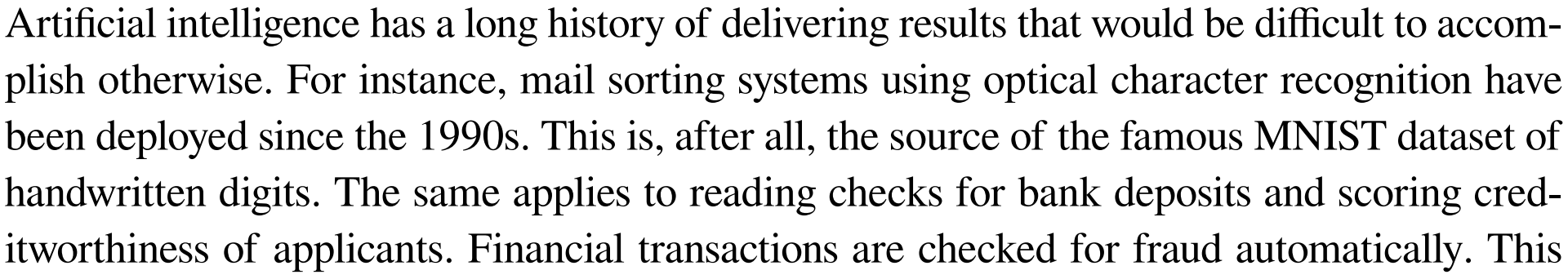

Artificial intelligence has a long history of delivering results that would be difficult to accomplish otherwise. For instance, mail sorting systems using optical character recognition have been deployed since the 1990s. This is, after all, the source of the famous MNIST dataset of handwritten digits. The same applies to reading checks for bank deposits and scoring creditworthiness of applicants. Financial transactions are checked for fraud automatically. This

42 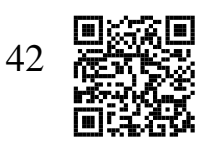

forms the backbone of many e-commerce payment systems, such as PayPal, Stripe, AliPay, WeChat, Apple, Visa, and MasterCard. Computer programs for chess have been competitive for decades. Machine learning feeds search, recommendation, personalization, and ranking on the Internet. In other words, machine learning is pervasive, albeit often hidden from sight.

It is only recently that AI has been in the limelight, mostly due to solutions to problems that were considered intractable previously and that are directly related to consumers. Many of such advances are attributed to deep learning.

- Intelligent assistants, such as Apple's Siri, Amazon's Alexa, and Google's assistant, are able to respond to spoken requests with a reasonable degree of accuracy. This includes menial jobs, like turning on light switches, and more complex tasks, such as arranging barber's appointments and offering phone support dialog. This is likely the most noticeable sign that AI is affecting our lives.

- A key ingredient in digital assistants is their ability to recognize speech accurately. The accuracy of such systems has gradually increased to the point of achieving parity with humans for certain applications (Xiong *et al.*, 2018).

- Object recognition has likewise come a long way. Identifying the object in a picture was a fairly challenging task in 2010. On the ImageNet benchmark researchers from NEC Labs and University of Illinois at Urbana-Champaign achieved a top-five error rate of 28% (Lin *et al.*, 2010). By 2017, this error rate was reduced to 2.25% (Hu *et al.*, 2018). Similarly, stunning results have been achieved for identifying birdsong and for diagnosing skin cancer.

- Prowess in games used to provide a measuring stick for human ability. Starting from TD-Gammon, a program for playing backgammon using temporal difference reinforcement learning, algorithmic and computational progress has led to algorithms for a wide range of applications. Compared with backgammon, chess has a much more complex state space and set of actions. DeepBlue beat Garry Kasparov using massive parallelism, special-purpose hardware and efficient search through the game tree (Campbell *et al.*, 2002). Go is more difficult still, due to its huge state space. AlphaGo reached human parity in 2015, using deep learning combined with Monte Carlo tree sampling (Silver *et al.*, 2016). The challenge in Poker was that the state space is large and only partially observed (we do not know the opponents' cards). Libratus exceeded human performance in Poker using efficiently structured strategies (Brown and Sandholm, 2017).

- Another indication of progress in AI is the advent of self-driving vehicles. While full autonomy is not yet within reach, excellent progress has been made in this direction, with companies such as Tesla, NVIDIA, and Waymo shipping products that enable partial autonomy. What makes full autonomy so challenging is that proper driving requires the ability to perceive, to reason and to incorporate rules into a system. At present, deep learning is used primarily in the visual aspect of these problems. The rest is heavily tuned by engineers.

This barely scratches the surface of significant applications of machine learning. For instance,

robotics, logistics, computational biology, particle physics, and astronomy owe some of their most impressive recent advances at least in parts to machine learning, which is thus becoming a ubiquitous tool for engineers and scientists.

Frequently, questions about a coming AI apocalypse and the plausibility of a *singularity* have been raised in non-technical articles. The fear is that somehow machine learning systems will become sentient and make decisions, independently of their programmers, that directly impact the lives of humans. To some extent, AI already affects the livelihood of humans in direct ways: creditworthiness is assessed automatically, autopilots mostly navigate vehicles, decisions about whether to grant bail use statistical data as input. More frivolously, we can ask Alexa to switch on the coffee machine.

Fortunately, we are far from a sentient AI system that could deliberately manipulate its human creators. First, AI systems are engineered, trained, and deployed in a specific, goal-oriented manner. While their behavior might give the illusion of general intelligence, it is a combination of rules, heuristics and statistical models that underlie the design. Second, at present, there are simply no tools for *artificial general intelligence* that are able to improve themselves, reason about themselves, and that are able to modify, extend, and improve their own architecture while trying to solve general tasks.

A much more pressing concern is how AI is being used in our daily lives. It is likely that many routine tasks, currently fulfilled by humans, can and will be automated. Farm robots will likely reduce the costs for organic farmers but they will also automate harvesting operations. This phase of the industrial revolution may have profound consequences for large swaths of society, since menial jobs provide much employment in many countries. Furthermore, statistical models, when applied without care, can lead to racial, gender, or age bias and raise reasonable concerns about procedural fairness if automated to drive consequential decisions. It is important to ensure that these algorithms are used with care. With what we know today, this strikes us as a much more pressing concern than the potential of malevolent superintelligence for destroying humanity.

## 1.7 The Essence of Deep Learning

Thus far, we have talked in broad terms about machine learning. Deep learning is the subset of machine learning concerned with models based on many-layered neural networks. It is *deep* in precisely the sense that its models learn many *layers* of transformations. While this might sound narrow, deep learning has given rise to a dizzying array of models, techniques, problem formulations, and applications. Many intuitions have been developed to explain the benefits of depth. Arguably, all machine learning has many layers of computation, the first consisting of feature processing steps. What differentiates deep learning is that the operations learned at each of the many layers of representations are learned jointly from data.

The problems that we have discussed so far, such as learning from the raw audio signal,

the raw pixel values of images, or mapping between sentences of arbitrary lengths and their counterparts in foreign languages, are those where deep learning excels and traditional methods falter. It turns out that these many-layered models are capable of addressing low-level perceptual data in a way that previous tools could not. Arguably the most significant commonality in deep learning methods is *end-to-end training*. That is, rather than assembling a system based on components that are individually tuned, one builds the system and then tunes their performance jointly. For instance, in computer vision scientists used to separate the process of *feature engineering* from the process of building machine learning models. The Canny edge detector (Canny, 1987) and Lowe's SIFT feature extractor (Lowe, 2004) reigned supreme for over a decade as algorithms for mapping images into feature vectors. In bygone days, the crucial part of applying machine learning to these problems consisted of coming up with manually-engineered ways of transforming the data into some form amenable to shallow models. Unfortunately, there is only so much that humans can accomplish by ingenuity in comparison with a consistent evaluation over millions of choices carried out automatically by an algorithm. When deep learning took over, these feature extractors were replaced by automatically tuned filters that yielded superior accuracy.

Thus, one key advantage of deep learning is that it replaces not only the shallow models at the end of traditional learning pipelines, but also the labor-intensive process of feature engineering. Moreover, by replacing much of the domain-specific preprocessing, deep learning has eliminated many of the boundaries that previously separated computer vision, speech recognition, natural language processing, medical informatics, and other application areas, thereby offering a unified set of tools for tackling diverse problems.

Beyond end-to-end training, we are experiencing a transition from parametric statistical descriptions to fully nonparametric models. When data is scarce, one needs to rely on simplifying assumptions about reality in order to obtain useful models. When data is abundant, these can be replaced by nonparametric models that better fit the data. To some extent, this mirrors the progress that physics experienced in the middle of the previous century with the availability of computers. Rather than solving by hand parametric approximations of how electrons behave, one can now resort to numerical simulations of the associated partial differential equations. This has led to much more accurate models, albeit often at the expense of interpretation.

Another difference from previous work is the acceptance of suboptimal solutions, dealing with nonconvex nonlinear optimization problems, and the willingness to try things before proving them. This new-found empiricism in dealing with statistical problems, combined with a rapid influx of talent has led to rapid progress in the development of practical algorithms, albeit in many cases at the expense of modifying and re-inventing tools that existed for decades.

In the end, the deep learning community prides itself on sharing tools across academic and corporate boundaries, releasing many excellent libraries, statistical models, and trained networks as open source. It is in this spirit that the notebooks forming this book are freely available for distribution and use. We have worked hard to lower the barriers of access for anyone wishing to learn about deep learning and we hope that our readers will benefit from this.

## 1.8 Summary

Machine learning studies how computer systems can leverage experience (often data) to improve performance at specific tasks. It combines ideas from statistics, data mining, and optimization. Often, it is used as a means of implementing AI solutions. As a class of machine learning, representational learning focuses on how to automatically find the appropriate way to represent data. Considered as multi-level representation learning through learning many layers of transformations, deep learning replaces not only the shallow models at the end of traditional machine learning pipelines, but also the labor-intensive process of feature engineering. Much of the recent progress in deep learning has been triggered by an abundance of data arising from cheap sensors and Internet-scale applications, and by significant progress in computation, mostly through GPUs. Furthermore, the availability of efficient deep learning frameworks has made design and implementation of whole system optimization significantly easier, and this is a key component in obtaining high performance.

## 1.9 Exercises

1. Which parts of code that you are currently writing could be "learned", i.e., improved by learning and automatically determining design choices that are made in your code? Does your code include heuristic design choices? What data might you need to learn the desired behavior?
2. Which problems that you encounter have many examples for their solution, yet no specific way for automating them? These may be prime candidates for using deep learning.
3. Describe the relationships between algorithms, data, and computation. How do characteristics of the data and the current available computational resources influence the appropriateness of various algorithms?
4. Name some settings where end-to-end training is not currently the default approach but where it might be useful.

Discussions[43].

43

# 2 Preliminaries

To prepare for your dive into deep learning, you will need a few survival skills: (i) techniques for storing and manipulating data; (ii) libraries for ingesting and preprocessing data from a variety of sources; (iii) knowledge of the basic linear algebraic operations that we apply to high-dimensional data elements; (iv) just enough calculus to determine which direction to adjust each parameter in order to decrease the loss function; (v) the ability to automatically compute derivatives so that you can forget much of the calculus you just learned; (vi) some basic fluency in probability, our primary language for reasoning under uncertainty; and (vii) some aptitude for finding answers in the official documentation when you get stuck.

In short, this chapter provides a rapid introduction to the basics that you will need to follow *most* of the technical content in this book.

## 2.1 Data Manipulation

In order to get anything done, we need some way to store and manipulate data. Generally, there are two important things we need to do with data: (i) acquire them; and (ii) process them once they are inside the computer. There is no point in acquiring data without some way to store it, so to start, let's get our hands dirty with $n$-dimensional arrays, which we also call *tensors*. If you already know the NumPy scientific computing package, this will be a breeze. For all modern deep learning frameworks, the *tensor class* (`ndarray` in MXNet, `Tensor` in PyTorch and TensorFlow) resembles NumPy's `ndarray`, with a few killer features added. First, the tensor class supports automatic differentiation. Second, it leverages GPUs to accelerate numerical computation, whereas NumPy only runs on CPUs. These properties make neural networks both easy to code and fast to run.

### 2.1.1 Getting Started

To start, we import the PyTorch library. Note that the package name is `torch`.

```
import torch
```

A tensor represents a (possibly multidimensional) array of numerical values. In the one-dimensional case, i.e., when only one axis is needed for the data, a tensor is called a *vector*.

With two axes, a tensor is called a *matrix*. With $k > 2$ axes, we drop the specialized names and just refer to the object as a $k^{\text{th}}$-*order tensor*.

PyTorch provides a variety of functions for creating new tensors prepopulated with values. For example, by invoking `arange(n)`, we can create a vector of evenly spaced values, starting at 0 (included) and ending at `n` (not included). By default, the interval size is 1. Unless otherwise specified, new tensors are stored in main memory and designated for CPU-based computation.

```
x = torch.arange(12, dtype=torch.float32)
x
```

```
tensor([ 0.,  1.,  2.,  3.,  4.,  5.,  6.,  7.,  8.,  9., 10., 11.])
```

Each of these values is called an *element* of the tensor. The tensor `x` contains 12 elements. We can inspect the total number of elements in a tensor via its `numel` method.

```
x.numel()
```

```
12
```

We can access a tensor's *shape* (the length along each axis) by inspecting its `shape` attribute. Because we are dealing with a vector here, the `shape` contains just a single element and is identical to the size.

```
x.shape
```

```
torch.Size([12])
```

We can change the shape of a tensor without altering its size or values, by invoking `reshape`. For example, we can transform our vector `x` whose shape is (12,) to a matrix `X` with shape (3, 4). This new tensor retains all elements but reconfigures them into a matrix. Notice that the elements of our vector are laid out one row at a time and thus `x[3] == X[0, 3]`.

```
X = x.reshape(3, 4)
X
```

```
tensor([[ 0.,  1.,  2.,  3.],
        [ 4.,  5.,  6.,  7.],
        [ 8.,  9., 10., 11.]])
```

Note that specifying every shape component to `reshape` is redundant. Because we already know our tensor's size, we can work out one component of the shape given the rest. For example, given a tensor of size $n$ and target shape $(h, w)$, we know that $w = n/h$. To automatically infer one component of the shape, we can place a `-1` for the shape component that should be inferred automatically. In our case, instead of calling `x.reshape(3, 4)`, we could have equivalently called `x.reshape(-1, 4)` or `x.reshape(3, -1)`.

Practitioners often need to work with tensors initialized to contain all 0s or 1s. We can construct a tensor with all elements set to 0 and a shape of (2, 3, 4) via the `zeros` function.

```
torch.zeros((2, 3, 4))
```

```
tensor([[[0., 0., 0., 0.],
         [0., 0., 0., 0.],
         [0., 0., 0., 0.]],

        [[0., 0., 0., 0.],
         [0., 0., 0., 0.],
         [0., 0., 0., 0.]]])
```

Similarly, we can create a tensor with all 1s by invoking `ones`.

```
torch.ones((2, 3, 4))
```

```
tensor([[[1., 1., 1., 1.],
         [1., 1., 1., 1.],
         [1., 1., 1., 1.]],

        [[1., 1., 1., 1.],
         [1., 1., 1., 1.],
         [1., 1., 1., 1.]]])
```

We often wish to sample each element randomly (and independently) from a given probability distribution. For example, the parameters of neural networks are often initialized randomly. The following snippet creates a tensor with elements drawn from a standard Gaussian (normal) distribution with mean 0 and standard deviation 1.

```
torch.randn(3, 4)
```

```
tensor([[ 0.3987,  0.6606,  1.1238, -0.2449],
        [-0.9343,  0.8246, -0.8358,  0.8182],
        [ 1.2168,  0.0476, -0.9897,  1.3701]])
```

Finally, we can construct tensors by supplying the exact values for each element by supplying (possibly nested) Python list(s) containing numerical literals. Here, we construct a matrix with a list of lists, where the outermost list corresponds to axis 0, and the inner list corresponds to axis 1.

```
torch.tensor([[2, 1, 4, 3], [1, 2, 3, 4], [4, 3, 2, 1]])
```

```
tensor([[2, 1, 4, 3],
        [1, 2, 3, 4],
        [4, 3, 2, 1]])
```

### 2.1.2 Indexing and Slicing

As with Python lists, we can access tensor elements by indexing (starting with 0). To access an element based on its position relative to the end of the list, we can use negative indexing. Finally, we can access whole ranges of indices via slicing (e.g., `X[start:stop]`), where the returned value includes the first index (`start`) *but not the last* (`stop`). Finally, when only one index (or slice) is specified for a $k^{\text{th}}$-order tensor, it is applied along axis 0. Thus, in the following code, `[-1]` selects the last row and `[1:3]` selects the second and third rows.

```
X[-1], X[1:3]
```

```
(tensor([ 8.,  9., 10., 11.]),
 tensor([[ 4.,  5.,  6.,  7.],
         [ 8.,  9., 10., 11.]]))
```

Beyond reading them, we can also *write* elements of a matrix by specifying indices.

```
X[1, 2] = 17
X
```

```
tensor([[ 0.,  1.,  2.,  3.],
        [ 4.,  5., 17.,  7.],
        [ 8.,  9., 10., 11.]])
```

If we want to assign multiple elements the same value, we apply the indexing on the left-hand side of the assignment operation. For instance, `[:2, :]` accesses the first and second rows, where `:` takes all the elements along axis 1 (column). While we discussed indexing for matrices, this also works for vectors and for tensors of more than two dimensions.

```
X[:2, :] = 12
X
```

```
tensor([[12., 12., 12., 12.],
        [12., 12., 12., 12.],
        [ 8.,  9., 10., 11.]])
```

### 2.1.3 Operations

Now that we know how to construct tensors and how to read from and write to their elements, we can begin to manipulate them with various mathematical operations. Among the most useful of these are the *elementwise* operations. These apply a standard scalar operation to each element of a tensor. For functions that take two tensors as inputs, elementwise operations apply some standard binary operator on each pair of corresponding elements. We can create an elementwise function from any function that maps from a scalar to a scalar.

In mathematical notation, we denote such *unary* scalar operators (taking one input) by the signature $f : \mathbb{R} \rightarrow \mathbb{R}$. This just means that the function maps from any real number onto some other real number. Most standard operators, including unary ones like $e^x$, can be applied elementwise.

```
torch.exp(x)
```

```
tensor([162754.7969, 162754.7969, 162754.7969, 162754.7969, 162754.7969,
        162754.7969, 162754.7969, 162754.7969,   2980.9580,   8103.0840,
         22026.4648,  59874.1406])
```

Likewise, we denote *binary* scalar operators, which map pairs of real numbers to a (single) real number via the signature $f : \mathbb{R}, \mathbb{R} \rightarrow \mathbb{R}$. Given any two vectors $\mathbf{u}$ and $\mathbf{v}$ *of the same shape*, and a binary operator $f$, we can produce a vector $\mathbf{c} = F(\mathbf{u}, \mathbf{v})$ by setting $c_i \leftarrow f(u_i, v_i)$ for all $i$, where $c_i$, $u_i$, and $v_i$ are the $i^\mathrm{th}$ elements of vectors $\mathbf{c}$, $\mathbf{u}$, and $\mathbf{v}$. Here, we produced the vector-valued $F : \mathbb{R}^d, \mathbb{R}^d \rightarrow \mathbb{R}^d$ by *lifting* the scalar function to an elementwise vector operation. The common standard arithmetic operators for addition (`+`), subtraction (`-`), multiplication (`*`), division (`/`), and exponentiation (`**`) have all been *lifted* to elementwise operations for identically-shaped tensors of arbitrary shape.

```
x = torch.tensor([1.0, 2, 4, 8])
y = torch.tensor([2, 2, 2, 2])
x + y, x - y, x * y, x / y, x ** y
```

```
(tensor([ 3.,  4.,  6., 10.]),
 tensor([-1.,  0.,  2.,  6.]),
 tensor([ 2.,  4.,  8., 16.]),
 tensor([0.5000, 1.0000, 2.0000, 4.0000]),
 tensor([ 1.,  4., 16., 64.]))
```

In addition to elementwise computations, we can also perform linear algebraic operations, such as dot products and matrix multiplications. We will elaborate on these in Section 2.3.

We can also *concatenate* multiple tensors, stacking them end-to-end to form a larger one. We just need to provide a list of tensors and tell the system along which axis to concatenate. The example below shows what happens when we concatenate two matrices along rows (axis 0) instead of columns (axis 1). We can see that the first output's axis-0 length (6) is the sum of the two input tensors' axis-0 lengths $(3 + 3)$; while the second output's axis-1 length (8) is the sum of the two input tensors' axis-1 lengths $(4 + 4)$.

```
X = torch.arange(12, dtype=torch.float32).reshape((3,4))
Y = torch.tensor([[2.0, 1, 4, 3], [1, 2, 3, 4], [4, 3, 2, 1]])
torch.cat((X, Y), dim=0), torch.cat((X, Y), dim=1)
```

```
(tensor([[ 0.,  1.,  2.,  3.],
         [ 4.,  5.,  6.,  7.],
         [ 8.,  9., 10., 11.],
         [ 2.,  1.,  4.,  3.],
         [ 1.,  2.,  3.,  4.],
         [ 4.,  3.,  2.,  1.]]),
 tensor([[ 0.,  1.,  2.,  3.,  2.,  1.,  4.,  3.],
         [ 4.,  5.,  6.,  7.,  1.,  2.,  3.,  4.],
         [ 8.,  9., 10., 11.,  4.,  3.,  2.,  1.]]))
```

Sometimes, we want to construct a binary tensor via *logical statements*. Take `X == Y` as an example. For each position `i, j`, if `X[i, j]` and `Y[i, j]` are equal, then the corresponding entry in the result takes value `1`, otherwise it takes value `0`.

```
X == Y
```

```
tensor([[False,  True, False,  True],
        [False, False, False, False],
        [False, False, False, False]])
```

Summing all the elements in the tensor yields a tensor with only one element.

```
X.sum()
```

```
tensor(66.)
```

### 2.1.4 Broadcasting

By now, you know how to perform elementwise binary operations on two tensors of the same shape. Under certain conditions, even when shapes differ, we can still perform elementwise binary operations by invoking the *broadcasting mechanism*. Broadcasting works according to the following two-step procedure: (i) expand one or both arrays by copying elements along axes with length 1 so that after this transformation, the two tensors have the same shape; (ii) perform an elementwise operation on the resulting arrays.

```
a = torch.arange(3).reshape((3, 1))
b = torch.arange(2).reshape((1, 2))
a, b
```

```
(tensor([[0],
         [1],
         [2]]),
 tensor([[0, 1]]))
```

Since `a` and `b` are $3 \times 1$ and $1 \times 2$ matrices, respectively, their shapes do not match up. Broadcasting produces a larger $3 \times 2$ matrix by replicating matrix `a` along the columns and matrix `b` along the rows before adding them elementwise.

```
a + b
```

```
tensor([[0, 1],
        [1, 2],
        [2, 3]])
```

### 2.1.5 Saving Memory

Running operations can cause new memory to be allocated to host results. For example, if we write `Y = X + Y`, we dereference the tensor that `Y` used to point to and instead point `Y` at the newly allocated memory. We can demonstrate this issue with Python's `id()` function, which gives us the exact address of the referenced object in memory. Note that after we run `Y = Y + X`, `id(Y)` points to a different location. That is because Python first evaluates `Y + X`, allocating new memory for the result and then points `Y` to this new location in memory.

```
before = id(Y)
Y = Y + X
id(Y) == before
```

```
False
```

This might be undesirable for two reasons. First, we do not want to run around allocating memory unnecessarily all the time. In machine learning, we often have hundreds of megabytes of parameters and update all of them multiple times per second. Whenever possible, we want to perform these updates *in place*. Second, we might point at the same parameters from multiple variables. If we do not update in place, we must be careful to update all of these references, lest we spring a memory leak or inadvertently refer to stale parameters.

Fortunately, performing in-place operations is easy. We can assign the result of an operation to a previously allocated array `Y` by using slice notation: `Y[:] = <expression>`. To illustrate this concept, we overwrite the values of tensor `Z`, after initializing it, using `zeros_like`, to have the same shape as `Y`.

```
Z = torch.zeros_like(Y)
print('id(Z):', id(Z))
Z[:] = X + Y
print('id(Z):', id(Z))
```

```
id(Z): 140496176783904
id(Z): 140496176783904
```

If the value of `X` is not reused in subsequent computations, we can also use `X[:] = X + Y` or `X += Y` to reduce the memory overhead of the operation.

```
before = id(X)
X += Y
id(X) == before
```

```
True
```

### 2.1.6 Conversion to Other Python Objects

Converting to a NumPy tensor (`ndarray`), or vice versa, is easy. The torch tensor and NumPy array will share their underlying memory, and changing one through an in-place operation will also change the other.

```
A = X.numpy()
B = torch.from_numpy(A)
type(A), type(B)
```

```
(numpy.ndarray, torch.Tensor)
```

To convert a size-1 tensor to a Python scalar, we can invoke the `item` function or Python's built-in functions.

```
a = torch.tensor([3.5])
a, a.item(), float(a), int(a)
```

```
(tensor([3.5000]), 3.5, 3.5, 3)
```

### 2.1.7 Summary

The tensor class is the main interface for storing and manipulating data in deep learning libraries. Tensors provide a variety of functionalities including construction routines; indexing and slicing; basic mathematics operations; broadcasting; memory-efficient assignment; and conversion to and from other Python objects.

### 2.1.8 Exercises

1. Run the code in this section. Change the conditional statement `X == Y` to `X < Y` or `X > Y`, and then see what kind of tensor you can get.
2. Replace the two tensors that operate by element in the broadcasting mechanism with other shapes, e.g., 3-dimensional tensors. Is the result the same as expected?

Discussions[44].

44

## 2.2 Data Preprocessing

So far, we have been working with synthetic data that arrived in ready-made tensors. However, to apply deep learning in the wild we must extract messy data stored in arbitrary formats, and preprocess it to suit our needs. Fortunately, the *pandas* library[45] can do much of the heavy lifting. This section, while no substitute for a proper *pandas* tutorial[46], will give you a crash course on some of the most common routines.

45 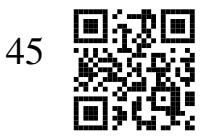

### 2.2.1 Reading the Dataset

Comma-separated values (CSV) files are ubiquitous for the storing of tabular (spreadsheet-like) data. In them, each line corresponds to one record and consists of several (comma-separated) fields, e.g., "Albert Einstein,March 14 1879,Ulm,Federal polytechnic school,field of gravitational physics". To demonstrate how to load CSV files with `pandas`, we create a CSV file below `../data/house_tiny.csv`. This file represents a dataset of homes, where each row corresponds to a distinct home and the columns correspond to the number of rooms (`NumRooms`), the roof type (`RoofType`), and the price (`Price`).

```
import os

os.makedirs(os.path.join('..', 'data'), exist_ok=True)
data_file = os.path.join('..', 'data', 'house_tiny.csv')
with open(data_file, 'w') as f:
    f.write('''NumRooms,RoofType,Price
NA,NA,127500
2,NA,106000
4,Slate,178100
NA,NA,140000''')
```

Now let's import `pandas` and load the dataset with `read_csv`.

```
import pandas as pd

data = pd.read_csv(data_file)
print(data)
```

```
   NumRooms RoofType   Price
0       NaN      NaN  127500
1       2.0      NaN  106000
2       4.0    Slate  178100
3       NaN      NaN  140000
```

### 2.2.2 Data Preparation

In supervised learning, we train models to predict a designated *target* value, given some set of *input* values. Our first step in processing the dataset is to separate out columns corresponding to input versus target values. We can select columns either by name or via integer-location based indexing (`iloc`).

You might have noticed that `pandas` replaced all CSV entries with value `NA` with a special `NaN` (*not a number*) value. This can also happen whenever an entry is empty, e.g., "3,,,270000". These are called *missing values* and they are the "bed bugs" of data science, a persistent menace that you will confront throughout your career. Depending upon the context, missing values might be handled either via *imputation* or *deletion*. Imputation replaces missing values with estimates of their values while deletion simply discards either those rows or those columns that contain missing values.

Here are some common imputation heuristics. For categorical input fields, we can treat `NaN` as a category. Since the `RoofType` column takes values `Slate` and `NaN`, `pandas` can convert this column into two columns `RoofType_Slate` and `RoofType_nan`. A row whose roof type is `Slate` will set values of `RoofType_Slate` and `RoofType_nan` to 1 and 0, respectively. The converse holds for a row with a missing `RoofType` value.

```
inputs, targets = data.iloc[:, 0:2], data.iloc[:, 2]
inputs = pd.get_dummies(inputs, dummy_na=True)
print(inputs)
```

```
   NumRooms  RoofType_Slate  RoofType_nan
0       NaN           False          True
1       2.0           False          True
2       4.0            True         False
3       NaN           False          True
```

For missing numerical values, one common heuristic is to replace the `NaN` entries with the mean value of the corresponding column.

```
inputs = inputs.fillna(inputs.mean())
print(inputs)
```

```
   NumRooms  RoofType_Slate  RoofType_nan
0       3.0           False          True
1       2.0           False          True
2       4.0            True         False
3       3.0           False          True
```

## 2.2.3 Conversion to the Tensor Format

Now that all the entries in `inputs` and `targets` are numerical, we can load them into a tensor (recall Section 2.1).

```
import torch

X = torch.tensor(inputs.to_numpy(dtype=float))
y = torch.tensor(targets.to_numpy(dtype=float))
X, y
```

```
(tensor([[3., 0., 1.],
         [2., 0., 1.],
         [4., 1., 0.],
         [3., 0., 1.]], dtype=torch.float64),
 tensor([127500., 106000., 178100., 140000.], dtype=torch.float64))
```

## 2.2.4 Discussion

You now know how to partition data columns, impute missing variables, and load `pandas` data into tensors. In Section 5.7, you will pick up some more data processing skills. While this crash course kept things simple, data processing can get hairy. For example, rather than arriving in a single CSV file, our dataset might be spread across multiple files extracted from a relational database. For instance, in an e-commerce application, customer addresses might live in one table and purchase data in another. Moreover, practitioners face myriad data types beyond categorical and numeric, for example, text strings, images, audio data, and point clouds. Oftentimes, advanced tools and efficient algorithms are required in order to prevent data processing from becoming the biggest bottleneck in the machine learning pipeline. These problems will arise when we get to computer vision and natural language processing. Finally, we must pay attention to data quality. Real-world datasets are often plagued by outliers, faulty measurements from sensors, and recording errors, which must be addressed before feeding the data into any model. Data visualization tools such as seaborn[47], Bokeh[48], or matplotlib[49] can help you to manually inspect the data and develop intuitions about the type of problems you may need to address.

47 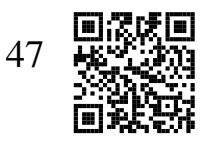

48 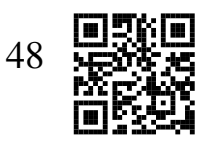

49 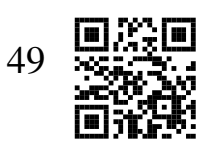

### 2.2.5 Exercises

1. Try loading datasets, e.g., Abalone from the UCI Machine Learning Repository[50] and inspect their properties. What fraction of them has missing values? What fraction of the variables is numerical, categorical, or text?

50 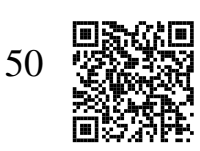

2. Try indexing and selecting data columns by name rather than by column number. The pandas documentation on indexing[51] has further details on how to do this.

51 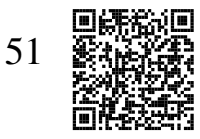

3. How large a dataset do you think you could load this way? What might be the limitations? Hint: consider the time to read the data, representation, processing, and memory footprint. Try this out on your laptop. What happens if you try it out on a server?
4. How would you deal with data that has a very large number of categories? What if the category labels are all unique? Should you include the latter?
5. What alternatives to pandas can you think of? How about loading NumPy tensors from a file[52]? Check out Pillow[53], the Python Imaging Library.

52 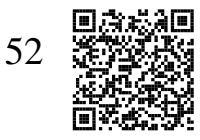

Discussions[54].

53 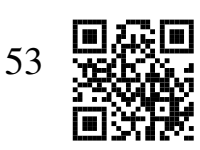

## 2.3 Linear Algebra

54

By now, we can load datasets into tensors and manipulate these tensors with basic mathematical operations. To start building sophisticated models, we will also need a few tools from linear algebra. This section offers a gentle introduction to the most essential concepts, starting from scalar arithmetic and ramping up to matrix multiplication.

```
import torch
```

### 2.3.1 Scalars

Most everyday mathematics consists of manipulating numbers one at a time. Formally, we call these values *scalars*. For example, the temperature in Palo Alto is a balmy 72 degrees Fahrenheit. If you wanted to convert the temperature to Celsius you would evaluate the expression $c = \frac{5}{9}(f - 32)$, setting $f$ to 72. In this equation, the values 5, 9, and 32 are constant scalars. The variables $c$ and $f$ in general represent unknown scalars.

We denote scalars by ordinary lower-cased letters (e.g., $x$, $y$, and $z$) and the space of all (continuous) *real-valued* scalars by $\mathbb{R}$. For expedience, we will skip past rigorous definitions of *spaces*: just remember that the expression $x \in \mathbb{R}$ is a formal way to say that $x$ is a real-valued scalar. The symbol $\in$ (pronounced "in") denotes membership in a set. For example, $x, y \in \{0, 1\}$ indicates that $x$ and $y$ are variables that can only take values 0 or 1.

Scalars are implemented as tensors that contain only one element. Below, we assign two scalars and perform the familiar addition, multiplication, division, and exponentiation operations.

```
x = torch.tensor(3.0)
y = torch.tensor(2.0)

x + y, x * y, x / y, x**y
```

```
(tensor(5.), tensor(6.), tensor(1.5000), tensor(9.))
```

### 2.3.2 Vectors

For current purposes, you can think of a vector as a fixed-length array of scalars. As with their code counterparts, we call these scalars the *elements* of the vector (synonyms include *entries* and *components*). When vectors represent examples from real-world datasets, their values hold some real-world significance. For example, if we were training a model to predict the risk of a loan defaulting, we might associate each applicant with a vector whose components correspond to quantities like their income, length of employment, or number of previous defaults. If we were studying the risk of heart attack, each vector might represent a patient and its components might correspond to their most recent vital signs, cholesterol levels, minutes of exercise per day, etc. We denote vectors by bold lowercase letters, (e.g., $\mathbf{x}$, $\mathbf{y}$, and $\mathbf{z}$).

Vectors are implemented as $1^{\textrm{st}}$-order tensors. In general, such tensors can have arbitrary lengths, subject to memory limitations. Caution: in Python, as in most programming languages, vector indices start at $0$, also known as *zero-based indexing*, whereas in linear algebra subscripts begin at $1$ (one-based indexing).

```
x = torch.arange(3)
x
```

```
tensor([0, 1, 2])
```

We can refer to an element of a vector by using a subscript. For example, $x_2$ denotes the second element of $\mathbf{x}$. Since $x_2$ is a scalar, we do not bold it. By default, we visualize vectors by stacking their elements vertically.

$$\mathbf{x} =\begin{bmatrix}x_{1} \\ \vdots \\ x_{n}\end{bmatrix}, \tag{2.3.1}$$

Here $x_1, \ldots, x_n$ are elements of the vector. Later on, we will distinguish between such *column vectors* and *row vectors* whose elements are stacked horizontally. Recall that we access a tensor's elements via indexing.

```
x[2]
```

```
tensor(2)
```

To indicate that a vector contains $n$ elements, we write $\mathbf{x} \in \mathbb{R}^n$. Formally, we call $n$ the *dimensionality* of the vector. In code, this corresponds to the tensor's length, accessible via Python's built-in `len` function.

```
len(x)
```

```
3
```

We can also access the length via the `shape` attribute. The shape is a tuple that indicates a tensor's length along each axis. Tensors with just one axis have shapes with just one element.

```
x.shape
```

```
torch.Size([3])
```

Oftentimes, the word "dimension" gets overloaded to mean both the number of axes and the length along a particular axis. To avoid this confusion, we use *order* to refer to the number of axes and *dimensionality* exclusively to refer to the number of components.

### 2.3.3 Matrices

Just as scalars are $0^{\textrm{th}}$-order tensors and vectors are $1^{\textrm{st}}$-order tensors, matrices are $2^{\textrm{nd}}$-order tensors. We denote matrices by bold capital letters (e.g., $\mathbf{X}$, $\mathbf{Y}$, and $\mathbf{Z}$), and represent them in code by tensors with two axes. The expression $\mathbf{A} \in \mathbb{R}^{m \times n}$ indicates that a matrix $\mathbf{A}$ contains $m \times n$ real-valued scalars, arranged as $m$ rows and $n$ columns. When $m = n$, we say that a matrix is *square*. Visually, we can illustrate any matrix as a table. To refer to an individual element, we subscript both the row and column indices, e.g., $a_{ij}$ is the value that belongs to $\mathbf{A}$'s $i^{\textrm{th}}$ row and $j^{\textrm{th}}$ column:

$$\mathbf{A}=\begin{bmatrix} a_{11} & a_{12} & \cdots & a_{1n} \\ a_{21} & a_{22} & \cdots & a_{2n} \\ \vdots & \vdots & \ddots & \vdots \\ a_{m1} & a_{m2} & \cdots & a_{mn} \\ \end{bmatrix}. \tag{2.3.2}$$

In code, we represent a matrix $\mathbf{A} \in \mathbb{R}^{m \times n}$ by a $2^{\textrm{nd}}$-order tensor with shape ($m$, $n$). We can convert any appropriately sized $m \times n$ tensor into an $m \times n$ matrix by passing the desired shape to `reshape`:

```
A = torch.arange(6).reshape(3, 2)
A
```

```
tensor([[0, 1],
        [2, 3],
        [4, 5]])
```

Sometimes we want to flip the axes. When we exchange a matrix's rows and columns, the result is called its *transpose*. Formally, we signify a matrix $\mathbf{A}$'s transpose by $\mathbf{A}^\top$ and if $\mathbf{B} = \mathbf{A}^\top$, then $b_{ij} = a_{ji}$ for all $i$ and $j$. Thus, the transpose of an $m \times n$ matrix is an $n \times m$ matrix:

$$\mathbf{A}^\top = \begin{bmatrix} a_{11} & a_{21} & \dots & a_{m1} \\ a_{12} & a_{22} & \dots & a_{m2} \\ \vdots & \vdots & \ddots & \vdots \\ a_{1n} & a_{2n} & \dots & a_{mn} \end{bmatrix}. \tag{2.3.3}$$

In code, we can access any matrix's transpose as follows:

```
A.T
```

```
tensor([[0, 2, 4],
        [1, 3, 5]])
```

Symmetric matrices are the subset of square matrices that are equal to their own transposes: $\mathbf{A} = \mathbf{A}^\top$. The following matrix is symmetric:

```
A = torch.tensor([[1, 2, 3], [2, 0, 4], [3, 4, 5]])
A == A.T
```

```
tensor([[True, True, True],
        [True, True, True],
        [True, True, True]])
```

Matrices are useful for representing datasets. Typically, rows correspond to individual records and columns correspond to distinct attributes.

### 2.3.4 Tensors

While you can go far in your machine learning journey with only scalars, vectors, and matrices, eventually you may need to work with higher-order tensors. Tensors give us a generic way of describing extensions to $n^{\text{th}}$-order arrays. We call software objects of the *tensor class* "tensors" precisely because they too can have arbitrary numbers of axes. While it may be confusing to use the word *tensor* for both the mathematical object and its realization in code, our meaning should usually be clear from context. We denote general tensors by capital letters with a special font face (e.g., $\mathsf{X}$, $\mathsf{Y}$, and $\mathsf{Z}$) and their indexing mechanism (e.g., $x_{ijk}$ and $[\mathsf{X}]_{1,2i-1,3}$) follows naturally from that of matrices.

Tensors will become more important when we start working with images. Each image arrives as a $3^{\textrm{rd}}$-order tensor with axes corresponding to the height, width, and *channel*. At each spatial location, the intensities of each color (red, green, and blue) are stacked along the channel. Furthermore, a collection of images is represented in code by a $4^{\textrm{th}}$-order tensor, where distinct images are indexed along the first axis. Higher-order tensors are constructed, as were vectors and matrices, by growing the number of shape components.

```
torch.arange(24).reshape(2, 3, 4)
```

```
tensor([[[ 0,  1,  2,  3],
         [ 4,  5,  6,  7],
         [ 8,  9, 10, 11]],

        [[12, 13, 14, 15],
         [16, 17, 18, 19],
         [20, 21, 22, 23]]])
```

## 2.3.5 Basic Properties of Tensor Arithmetic

Scalars, vectors, matrices, and higher-order tensors all have some handy properties. For example, elementwise operations produce outputs that have the same shape as their operands.

```
A = torch.arange(6, dtype=torch.float32).reshape(2, 3)
B = A.clone()  # Assign a copy of A to B by allocating new memory
A, A + B
```

```
(tensor([[0., 1., 2.],
         [3., 4., 5.]]),
 tensor([[ 0.,  2.,  4.],
         [ 6.,  8., 10.]]))
```

The elementwise product of two matrices is called their *Hadamard product* (denoted $\odot$). We can spell out the entries of the Hadamard product of two matrices $\mathbf{A}, \mathbf{B} \in \mathbb{R}^{m \times n}$:

$$\mathbf{A} \odot \mathbf{B} = \begin{bmatrix} a_{11} b_{11} & a_{12} b_{12} & \dots & a_{1n} b_{1n} \\ a_{21} b_{21} & a_{22} b_{22} & \dots & a_{2n} b_{2n} \\ \vdots & \vdots & \ddots & \vdots \\ a_{m1} b_{m1} & a_{m2} b_{m2} & \dots & a_{mn} b_{mn} \end{bmatrix}. \tag{2.3.4}$$

```
A * B
```

```
tensor([[ 0.,  1.,  4.],
        [ 9., 16., 25.]])
```

Adding or multiplying a scalar and a tensor produces a result with the same shape as the original tensor. Here, each element of the tensor is added to (or multiplied by) the scalar.

```
a = 2
X = torch.arange(24).reshape(2, 3, 4)
a + X, (a * X).shape
```

```
(tensor([[[ 2,  3,  4,  5],
          [ 6,  7,  8,  9],
          [10, 11, 12, 13]],

         [[14, 15, 16, 17],
          [18, 19, 20, 21],
          [22, 23, 24, 25]]]),
 torch.Size([2, 3, 4]))
```

### 2.3.6 Reduction

Often, we wish to calculate the sum of a tensor's elements. To express the sum of the elements in a vector $\mathbf{x}$ of length $n$, we write $\sum_{i=1}^{n} x_i$. There is a simple function for it:

```
x = torch.arange(3, dtype=torch.float32)
x, x.sum()
```

```
(tensor([0., 1., 2.]), tensor(3.))
```

To express sums over the elements of tensors of arbitrary shape, we simply sum over all its axes. For example, the sum of the elements of an $m \times n$ matrix $\mathbf{A}$ could be written $\sum_{i=1}^{m} \sum_{j=1}^{n} a_{ij}$.

```
A.shape, A.sum()
```

```
(torch.Size([2, 3]), tensor(15.))
```

By default, invoking the sum function *reduces* a tensor along all of its axes, eventually producing a scalar. Our libraries also allow us to specify the axes along which the tensor should be reduced. To sum over all elements along the rows (axis 0), we specify `axis=0` in `sum`. Since the input matrix reduces along axis 0 to generate the output vector, this axis is missing from the shape of the output.

```
A.shape, A.sum(axis=0).shape
```

```
(torch.Size([2, 3]), torch.Size([3]))
```

Specifying `axis=1` will reduce the column dimension (axis 1) by summing up elements of all the columns.

```
A.shape, A.sum(axis=1).shape
```

```
(torch.Size([2, 3]), torch.Size([2]))
```

Reducing a matrix along both rows and columns via summation is equivalent to summing up all the elements of the matrix.

```
A.sum(axis=[0, 1]) == A.sum()  # Same as A.sum()
```

```
tensor(True)
```

A related quantity is the *mean*, also called the *average*. We calculate the mean by dividing the sum by the total number of elements. Because computing the mean is so common, it gets a dedicated library function that works analogously to `sum`.

```
A.mean(), A.sum() / A.numel()
```

```
(tensor(2.5000), tensor(2.5000))
```

Likewise, the function for calculating the mean can also reduce a tensor along specific axes.

```
A.mean(axis=0), A.sum(axis=0) / A.shape[0]
```

```
(tensor([1.5000, 2.5000, 3.5000]), tensor([1.5000, 2.5000, 3.5000]))
```

### 2.3.7 Non-Reduction Sum

Sometimes it can be useful to keep the number of axes unchanged when invoking the function for calculating the sum or mean. This matters when we want to use the broadcast mechanism.

```
sum_A = A.sum(axis=1, keepdims=True)
sum_A, sum_A.shape
```

```
(tensor([[ 3.],
         [12.]]),
 torch.Size([2, 1]))
```

For instance, since `sum_A` keeps its two axes after summing each row, we can divide `A` by `sum_A` with broadcasting to create a matrix where each row sums up to 1.

```
A / sum_A
```

```
tensor([[0.0000, 0.3333, 0.6667],
        [0.2500, 0.3333, 0.4167]])
```

If we want to calculate the cumulative sum of elements of `A` along some axis, say `axis=0` (row by row), we can call the `cumsum` function. By design, this function does not reduce the input tensor along any axis.

```
A.cumsum(axis=0)
```

```
tensor([[0., 1., 2.],
        [3., 5., 7.]])
```

### 2.3.8 Dot Products

So far, we have only performed elementwise operations, sums, and averages. And if this was all we could do, linear algebra would not deserve its own section. Fortunately, this is where things get more interesting. One of the most fundamental operations is the dot product. Given two vectors $\mathbf{x}, \mathbf{y} \in \mathbb{R}^d$, their *dot product* $\mathbf{x}^\top \mathbf{y}$ (also known as *inner product*, $\langle \mathbf{x}, \mathbf{y} \rangle$) is a sum over the products of the elements at the same position: $\mathbf{x}^\top \mathbf{y} = \sum_{i=1}^{d} x_i y_i$.

```
y = torch.ones(3, dtype = torch.float32)
x, y, torch.dot(x, y)
```

```
(tensor([0., 1., 2.]), tensor([1., 1., 1.]), tensor(3.))
```

Equivalently, we can calculate the dot product of two vectors by performing an elementwise multiplication followed by a sum:

```
torch.sum(x * y)
```

```
tensor(3.)
```

Dot products are useful in a wide range of contexts. For example, given some set of values,

denoted by a vector $\mathbf{x} \in \mathbb{R}^n$, and a set of weights, denoted by $\mathbf{w} \in \mathbb{R}^n$, the weighted sum of the values in $\mathbf{x}$ according to the weights $\mathbf{w}$ could be expressed as the dot product $\mathbf{x}^\top \mathbf{w}$. When the weights are nonnegative and sum to 1, i.e., $\left(\sum_{i=1}^{n} w_i = 1\right)$, the dot product expresses a *weighted average*. After normalizing two vectors to have unit length, the dot products express the cosine of the angle between them. Later in this section, we will formally introduce this notion of *length*.

### 2.3.9 Matrix–Vector Products

Now that we know how to calculate dot products, we can begin to understand the *product* between an $m \times n$ matrix $\mathbf{A}$ and an $n$-dimensional vector $\mathbf{x}$. To start off, we visualize our matrix in terms of its row vectors

$$\mathbf{A} = \begin{bmatrix} \mathbf{a}^\top_{1} \\ \mathbf{a}^\top_{2} \\ \vdots \\ \mathbf{a}^\top_m \end{bmatrix}, \tag{2.3.5}$$

where each $\mathbf{a}^\top_{i} \in \mathbb{R}^n$ is a row vector representing the $i^\textrm{th}$ row of the matrix $\mathbf{A}$.

The matrix–vector product $\mathbf{A}\mathbf{x}$ is simply a column vector of length $m$, whose $i^\textrm{th}$ element is the dot product $\mathbf{a}^\top_i \mathbf{x}$:

$$\mathbf{A}\mathbf{x} = \begin{bmatrix} \mathbf{a}^\top_{1} \\ \mathbf{a}^\top_{2} \\ \vdots \\ \mathbf{a}^\top_m \end{bmatrix}\mathbf{x} = \begin{bmatrix} \mathbf{a}^\top_{1} \mathbf{x} \\ \mathbf{a}^\top_{2} \mathbf{x} \\ \vdots\\ \mathbf{a}^\top_{m} \mathbf{x}\\ \end{bmatrix}. \tag{2.3.6}$$

We can think of multiplication with a matrix $\mathbf{A}\in \mathbb{R}^{m \times n}$ as a transformation that projects vectors from $\mathbb{R}^{n}$ to $\mathbb{R}^{m}$. These transformations are remarkably useful. For example, we can represent rotations as multiplications by certain square matrices. Matrix–vector products also describe the key calculation involved in computing the outputs of each layer in a neural network given the outputs from the previous layer.

To express a matrix–vector product in code, we use the `mv` function. Note that the column dimension of `A` (its length along axis 1) must be the same as the dimension of `x` (its length). Python has a convenience operator `@` that can execute both matrix–vector and matrix–matrix products (depending on its arguments). Thus we can write `A@x`.

```
A.shape, x.shape, torch.mv(A, x), A@x
```

```
(torch.Size([2, 3]), torch.Size([3]), tensor([ 5., 14.]), tensor([ 5., 14.]))
```

### 2.3.10 Matrix–Matrix Multiplication

Once you have gotten the hang of dot products and matrix–vector products, then *matrix–matrix multiplication* should be straightforward.

Say that we have two matrices $\mathbf{A} \in \mathbb{R}^{n\times k}$ and $\mathbf{B} \in \mathbb{R}^{k\times m}$:

$$\mathbf{A}=\begin{bmatrix} a_{11} & a_{12} & \cdots & a_{1k} \\ a_{21} & a_{22} & \cdots & a_{2k} \\ \vdots & \vdots & \ddots & \vdots \\ a_{n1} & a_{n2} & \cdots & a_{nk} \end{bmatrix},\quad \mathbf{B}=\begin{bmatrix} b_{11} & b_{12} & \cdots & b_{1m} \\ b_{21} & b_{22} & \cdots & b_{2m} \\ \vdots & \vdots & \ddots & \vdots \\ b_{k1} & b_{k2} & \cdots & b_{km} \end{bmatrix}. \tag{2.3.7}$$

Let $\mathbf{a}^\top_i \in \mathbb{R}^k$ denote the row vector representing the $i^\textrm{th}$ row of the matrix $\mathbf{A}$ and let $\mathbf{b}_j \in \mathbb{R}^k$ denote the column vector from the $j^\textrm{th}$ column of the matrix $\mathbf{B}$:

$$\mathbf{A}= \begin{bmatrix} \mathbf{a}^\top_{1} \\ \mathbf{a}^\top_{2} \\ \vdots \\ \mathbf{a}^\top_n \end{bmatrix}, \quad \mathbf{B}=\begin{bmatrix} \mathbf{b}_{1} & \mathbf{b}_{2} & \cdots & \mathbf{b}_{m} \end{bmatrix}. \tag{2.3.8}$$

To form the matrix product $\mathbf{C} \in \mathbb{R}^{n\times m}$, we simply compute each element $c_{ij}$ as the dot product between the $i^\textrm{th}$ row of $\mathbf{A}$ and the $j^\textrm{th}$ column of $\mathbf{B}$, i.e., $\mathbf{a}^\top_i \mathbf{b}_j$:

$$\mathbf{C} = \mathbf{AB} = \begin{bmatrix} \mathbf{a}^\top_{1} \\ \mathbf{a}^\top_{2} \\ \vdots \\ \mathbf{a}^\top_n \end{bmatrix} \begin{bmatrix} \mathbf{b}_{1} & \mathbf{b}_{2} & \cdots & \mathbf{b}_{m} \end{bmatrix} = \begin{bmatrix} \mathbf{a}^\top_{1} \mathbf{b}_1 & \mathbf{a}^\top_{1}\mathbf{b}_2& \cdots & \mathbf{a}^\top_{1} \mathbf{b}_m \\ \mathbf{a}^\top_{2}\mathbf{b}_1 & \mathbf{a}^\top_{2} \mathbf{b}_2 & \cdots & \mathbf{a}^\top_{2} \mathbf{b}_m \\ \vdots & \vdots & \ddots &\vdots\\ \mathbf{a}^\top_{n} \mathbf{b}_1 & \mathbf{a}^\top_{n}\mathbf{b}_2& \cdots& \mathbf{a}^\top_{n} \mathbf{b}_m \end{bmatrix}. \tag{2.3.9}$$

We can think of the matrix–matrix multiplication $\mathbf{AB}$ as performing $m$ matrix–vector products or $m \times n$ dot products and stitching the results together to form an $n \times m$ matrix. In the following snippet, we perform matrix multiplication on `A` and `B`. Here, `A` is a matrix with two rows and three columns, and `B` is a matrix with three rows and four columns. After multiplication, we obtain a matrix with two rows and four columns.

```
B = torch.ones(3, 4)
torch.mm(A, B), A@B
```

```
(tensor([[ 3.,  3.,  3.,  3.],
         [12., 12., 12., 12.]]),
 tensor([[ 3.,  3.,  3.,  3.],
         [12., 12., 12., 12.]]))
```

The term *matrix–matrix multiplication* is often simplified to *matrix multiplication*, and should not be confused with the Hadamard product.

## 2.3.11 Norms

Some of the most useful operators in linear algebra are *norms*. Informally, the norm of a vector tells us how *big* it is. For instance, the $\ell_2$ norm measures the (Euclidean) length of a vector. Here, we are employing a notion of *size* that concerns the magnitude of a vector's components (not its dimensionality).

A norm is a function $\|\cdot\|$ that maps a vector to a scalar and satisfies the following three properties:

1. Given any vector $\mathbf{x}$, if we scale (all elements of) the vector by a scalar $\alpha \in \mathbb{R}$, its norm scales accordingly:

$$\|\alpha \mathbf{x}\| = |\alpha| \|\mathbf{x}\|. \tag{2.3.10}$$

2. For any vectors $\mathbf{x}$ and $\mathbf{y}$: norms satisfy the triangle inequality:

$$\|\mathbf{x} + \mathbf{y}\| \leq \|\mathbf{x}\| + \|\mathbf{y}\|. \tag{2.3.11}$$

3. The norm of a vector is nonnegative and it only vanishes if the vector is zero:

$$\|\mathbf{x}\| > 0 \text{ for all } \mathbf{x} \neq 0. \tag{2.3.12}$$

Many functions are valid norms and different norms encode different notions of size. The Euclidean norm that we all learned in elementary school geometry when calculating the hypotenuse of a right triangle is the square root of the sum of squares of a vector's elements. Formally, this is called the $\ell_2$ *norm* and expressed as

$$\|\mathbf{x}\|_2 = \sqrt{\sum_{i=1}^n x_i^2}. \tag{2.3.13}$$

The method `norm` calculates the $\ell_2$ norm.

```
u = torch.tensor([3.0, -4.0])
torch.norm(u)
```

```
tensor(5.)
```

The $\ell_1$ norm is also common and the associated measure is called the Manhattan distance. By definition, the $\ell_1$ norm sums the absolute values of a vector's elements:

$$\|\mathbf{x}\|_1 = \sum_{i=1}^n |x_i|. \tag{2.3.14}$$

Compared to the $\ell_2$ norm, it is less sensitive to outliers. To compute the $\ell_1$ norm, we compose the absolute value with the sum operation.

```
torch.abs(u).sum()
```

```
tensor(7.)
```

Both the $\ell_2$ and $\ell_1$ norms are special cases of the more general $\ell_p$ *norms*:

$$\|\mathbf{x}\|_p = \left(\sum_{i=1}^n |x_i|^p\right)^{1/p}. \tag{2.3.15}$$

In the case of matrices, matters are more complicated. After all, matrices can be viewed both as collections of individual entries *and* as objects that operate on vectors and transform them into other vectors. For instance, we can ask by how much longer the matrix–vector product $\mathbf{X}\mathbf{v}$ could be relative to $\mathbf{v}$. This line of thought leads to what is called the *spectral* norm. For now, we introduce the *Frobenius norm*, which is much easier to compute and defined as the square root of the sum of the squares of a matrix's elements:

$$\|\mathbf{X}\|_\textrm{F} = \sqrt{\sum_{i=1}^m \sum_{j=1}^n x_{ij}^2}. \tag{2.3.16}$$

The Frobenius norm behaves as if it were an $\ell_2$ norm of a matrix-shaped vector. Invoking the following function will calculate the Frobenius norm of a matrix.

```
torch.norm(torch.ones((4, 9)))
```

```
tensor(6.)
```

While we do not want to get too far ahead of ourselves, we already can plant some intuition about why these concepts are useful. In deep learning, we are often trying to solve optimization problems: *maximize* the probability assigned to observed data; *maximize* the revenue associated with a recommender model; *minimize* the distance between predictions and the ground truth observations; *minimize* the distance between representations of photos of the same person while *maximizing* the distance between representations of photos of different people. These distances, which constitute the objectives of deep learning algorithms, are often expressed as norms.

## 2.3.12 Discussion

In this section, we have reviewed all the linear algebra that you will need to understand a significant chunk of modern deep learning. There is a lot more to linear algebra, though, and much of it is useful for machine learning. For example, matrices can be decomposed into factors, and these decompositions can reveal low-dimensional structure in real-world datasets. There are entire subfields of machine learning that focus on using matrix decompositions and their generalizations to high-order tensors to discover structure in datasets and solve prediction problems. But this book focuses on deep learning. And we believe you will be more inclined to learn more mathematics once you have gotten your hands dirty applying machine learning to real datasets. So while we reserve the right to introduce more mathematics later on, we wrap up this section here.

If you are eager to learn more linear algebra, there are many excellent books and online resources. For a more advanced crash course, consider checking out Strang (1993), Kolter (2008), and Petersen and Pedersen (2008).

To recap:

- Scalars, vectors, matrices, and tensors are the basic mathematical objects used in linear algebra and have zero, one, two, and an arbitrary number of axes, respectively.
- Tensors can be sliced or reduced along specified axes via indexing, or operations such as `sum` and `mean`, respectively.
- Elementwise products are called Hadamard products. By contrast, dot products, matrix–vector products, and matrix–matrix products are not elementwise operations and in general return objects having shapes that are different from the the operands.
- Compared to Hadamard products, matrix–matrix products take considerably longer to compute (cubic rather than quadratic time).
- Norms capture various notions of the magnitude of a vector (or matrix), and are commonly applied to the difference of two vectors to measure their distance apart.
- Common vector norms include the $\ell_1$ and $\ell_2$ norms, and common matrix norms include the *spectral* and *Frobenius* norms.

### 2.3.13 Exercises

1. Prove that the transpose of the transpose of a matrix is the matrix itself: $(\mathbf{A}^\top)^\top = \mathbf{A}$.
2. Given two matrices $\mathbf{A}$ and $\mathbf{B}$, show that sum and transposition commute: $\mathbf{A}^\top + \mathbf{B}^\top = (\mathbf{A} + \mathbf{B})^\top$.
3. Given any square matrix $\mathbf{A}$, is $\mathbf{A} + \mathbf{A}^\top$ always symmetric? Can you prove the result by using only the results of the previous two exercises?
4. We defined the tensor `X` of shape (2, 3, 4) in this section. What is the output of `len(X)`? Write your answer without implementing any code, then check your answer using code.
5. For a tensor `X` of arbitrary shape, does `len(X)` always correspond to the length of a certain axis of `X`? What is that axis?
6. Run `A / A.sum(axis=1)` and see what happens. Can you analyze the results?
7. When traveling between two points in downtown Manhattan, what is the distance that you need to cover in terms of the coordinates, i.e., in terms of avenues and streets? Can you travel diagonally?
8. Consider a tensor of shape (2, 3, 4). What are the shapes of the summation outputs along axes 0, 1, and 2?
9. Feed a tensor with three or more axes to the `linalg.norm` function and observe its output. What does this function compute for tensors of arbitrary shape?

10. Consider three large matrices, say $\mathbf{A} \in \mathbb{R}^{2^{10} \times 2^{16}}$, $\mathbf{B} \in \mathbb{R}^{2^{16} \times 2^{5}}$ and $\mathbf{C} \in \mathbb{R}^{2^{5} \times 2^{14}}$, initialized with Gaussian random variables. You want to compute the product $\mathbf{ABC}$. Is there any difference in memory footprint and speed, depending on whether you compute $(\mathbf{AB})\mathbf{C}$ or $\mathbf{A}(\mathbf{BC})$. Why?

11. Consider three large matrices, say $\mathbf{A} \in \mathbb{R}^{2^{10} \times 2^{16}}$, $\mathbf{B} \in \mathbb{R}^{2^{16} \times 2^{5}}$ and $\mathbf{C} \in \mathbb{R}^{2^{5} \times 2^{16}}$. Is there any difference in speed depending on whether you compute $\mathbf{AB}$ or $\mathbf{AC}^\top$? Why? What changes if you initialize $\mathbf{C} = \mathbf{B}^\top$ without cloning memory? Why?

12. Consider three matrices, say $\mathbf{A}, \mathbf{B}, \mathbf{C} \in \mathbb{R}^{100 \times 200}$. Construct a tensor with three axes by stacking $[\mathbf{A}, \mathbf{B}, \mathbf{C}]$. What is the dimensionality? Slice out the second coordinate of the third axis to recover $\mathbf{B}$. Check that your answer is correct.

55 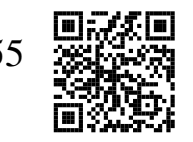 Discussions[55].

## 2.4 Calculus

For a long time, how to calculate the area of a circle remained a mystery. Then, in Ancient Greece, the mathematician Archimedes came up with the clever idea to inscribe a series of polygons with increasing numbers of vertices on the inside of a circle (Fig. 2.4.1). For a polygon with $n$ vertices, we obtain $n$ triangles. The height of each triangle approaches the radius $r$ as we partition the circle more finely. At the same time, its base approaches $2\pi r/n$, since the ratio between arc and secant approaches 1 for a large number of vertices. Thus, the area of the polygon approaches $n \cdot r \cdot \frac{1}{2}(2\pi r/n) = \pi r^2$.

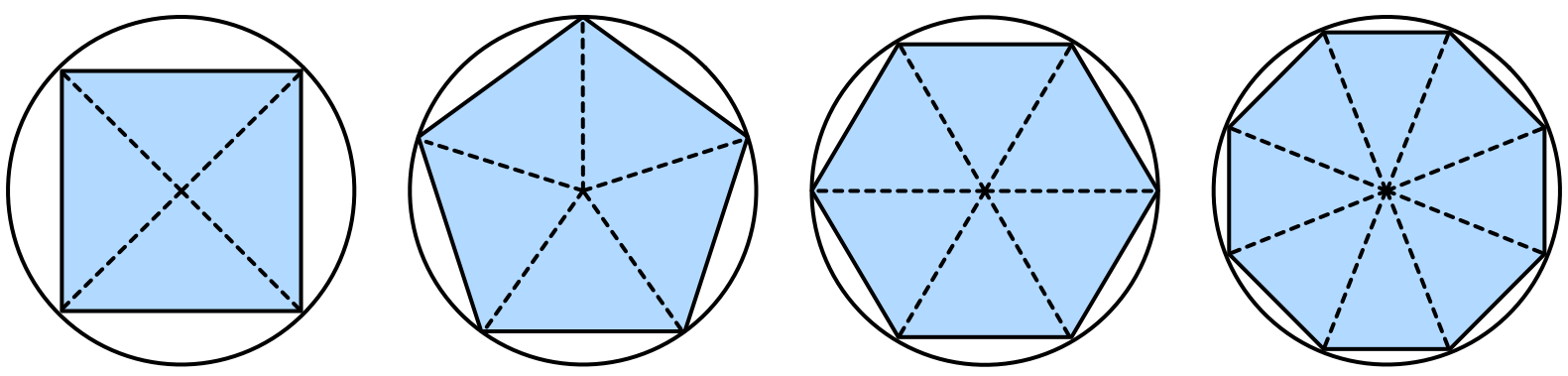

Fig. 2.4.1 Finding the area of a circle as a limit procedure.

This limiting procedure is at the root of both *differential calculus* and *integral calculus*. The former can tell us how to increase or decrease a function's value by manipulating its arguments. This comes in handy for the *optimization problems* that we face in deep learning, where we repeatedly update our parameters in order to decrease the loss function. Optimization addresses how to fit our models to training data, and calculus is its key prerequisite. However, do not forget that our ultimate goal is to perform well on *previously unseen* data. That problem is called *generalization* and will be a key focus of other chapters.

```
%matplotlib inline
import numpy as np
from matplotlib_inline import backend_inline
from d2l import torch as d2l
```

### 2.4.1 Derivatives and Differentiation

Put simply, a *derivative* is the rate of change in a function with respect to changes in its arguments. Derivatives can tell us how rapidly a loss function would increase or decrease were we to *increase* or *decrease* each parameter by an infinitesimally small amount. Formally, for functions $f : \mathbb{R} \to \mathbb{R}$, that map from scalars to scalars, the *derivative* of $f$ at a point $x$ is defined as

$$f'(x) = \lim_{h \to 0} \frac{f(x+h) - f(x)}{h}. \tag{2.4.1}$$

This term on the right hand side is called a *limit* and it tells us what happens to the value of an expression as a specified variable approaches a particular value. This limit tells us what the ratio between a perturbation $h$ and the change in the function value $f(x+h) - f(x)$ converges to as we shrink its size to zero.

When $f'(x)$ exists, $f$ is said to be *differentiable* at $x$; and when $f'(x)$ exists for all $x$ on a set, e.g., the interval $[a, b]$, we say that $f$ is differentiable on this set. Not all functions are differentiable, including many that we wish to optimize, such as accuracy and the area under the receiving operating characteristic (AUC). However, because computing the derivative of the loss is a crucial step in nearly all algorithms for training deep neural networks, we often optimize a differentiable *surrogate* instead.

We can interpret the derivative $f'(x)$ as the *instantaneous* rate of change of $f(x)$ with respect to $x$. Let's develop some intuition with an example. Define $u = f(x) = 3x^2 - 4x$.

```
def f(x):
    return 3 * x ** 2 - 4 * x
```

Setting $x = 1$, we see that $\frac{f(x+h)-f(x)}{h}$ approaches 2 as $h$ approaches 0. While this experiment lacks the rigor of a mathematical proof, we can quickly see that indeed $f'(1) = 2$.

```
for h in 10.0**np.arange(-1, -6, -1):
    print(f'h={h:.5f}, numerical limit={(f(1+h)-f(1))/h:.5f}')
```

```
h=0.10000, numerical limit=2.30000
h=0.01000, numerical limit=2.03000
h=0.00100, numerical limit=2.00300
h=0.00010, numerical limit=2.00030
h=0.00001, numerical limit=2.00003
```

There are several equivalent notational conventions for derivatives. Given $y = f(x)$, the following expressions are equivalent:

$$f'(x) = y' = \frac{dy}{dx} = \frac{df}{dx} = \frac{d}{dx} f(x) = Df(x) = D_x f(x), \tag{2.4.2}$$

where the symbols $\frac{d}{dx}$ and $D$ are *differentiation operators*. Below, we present the derivatives of some common functions:

$$\begin{aligned} \frac{d}{dx} C &= 0 && \text{for any constant } C \\ \frac{d}{dx} x^n &= n x^{n-1} && \text{for } n \neq 0 \\ \frac{d}{dx} e^x &= e^x \\ \frac{d}{dx} \ln x &= x^{-1}. \end{aligned} \tag{2.4.3}$$

Functions composed from differentiable functions are often themselves differentiable. The following rules come in handy for working with compositions of any differentiable functions $f$ and $g$, and constant $C$.

$$\begin{aligned} \frac{d}{dx} [C f(x)] &= C \frac{d}{dx} f(x) && \text{Constant multiple rule} \\ \frac{d}{dx} [f(x) + g(x)] &= \frac{d}{dx} f(x) + \frac{d}{dx} g(x) && \text{Sum rule} \\ \frac{d}{dx} [f(x) g(x)] &= f(x) \frac{d}{dx} g(x) + g(x) \frac{d}{dx} f(x) && \text{Product rule} \\ \frac{d}{dx} \frac{f(x)}{g(x)} &= \frac{g(x) \frac{d}{dx} f(x) - f(x) \frac{d}{dx} g(x)}{g^2(x)} && \text{Quotient rule} \end{aligned} \tag{2.4.4}$$

Using this, we can apply the rules to find the derivative of $3x^2 - 4x$ via

$$\frac{d}{dx} [3x^2 - 4x] = 3 \frac{d}{dx} x^2 - 4 \frac{d}{dx} x = 6x - 4. \tag{2.4.5}$$

Plugging in $x = 1$ shows that, indeed, the derivative equals 2 at this location. Note that derivatives tell us the *slope* of a function at a particular location.

### 2.4.2 Visualization Utilities

We can visualize the slopes of functions using the `matplotlib` library. We need to define a few functions. As its name indicates, `use_svg_display` tells `matplotlib` to output graphics in SVG format for crisper images. The comment `#@save` is a special modifier that allows us to save any function, class, or other code block to the `d2l` package so that we can invoke it later without repeating the code, e.g., via `d2l.use_svg_display()`.

```
def use_svg_display():  #@save
    """Use the svg format to display a plot in Jupyter."""
    backend_inline.set_matplotlib_formats('svg')
```

Conveniently, we can set figure sizes with `set_figsize`. Since the import statement `from matplotlib import pyplot as plt` was marked via `#@save` in the `d2l` package, we can call `d2l.plt`.

```
def set_figsize(figsize=(3.5, 2.5)):  #@save
    """Set the figure size for matplotlib."""
    use_svg_display()
    d2l.plt.rcParams['figure.figsize'] = figsize
```

The `set_axes` function can associate axes with properties, including labels, ranges, and scales.

```
#@save
def set_axes(axes, xlabel, ylabel, xlim, ylim, xscale, yscale, legend):
    """Set the axes for matplotlib."""
    axes.set_xlabel(xlabel), axes.set_ylabel(ylabel)
    axes.set_xscale(xscale), axes.set_yscale(yscale)
    axes.set_xlim(xlim),     axes.set_ylim(ylim)
    if legend:
        axes.legend(legend)
    axes.grid()
```

With these three functions, we can define a `plot` function to overlay multiple curves. Much of the code here is just ensuring that the sizes and shapes of inputs match.

```
#@save
def plot(X, Y=None, xlabel=None, ylabel=None, legend=[], xlim=None,
         ylim=None, xscale='linear', yscale='linear',
         fmts=('-', 'm--', 'g-.', 'r:'), figsize=(3.5, 2.5), axes=None):
    """Plot data points."""

    def has_one_axis(X):  # True if X (tensor or list) has 1 axis
        return (hasattr(X, "ndim") and X.ndim == 1 or isinstance(X, list)
                and not hasattr(X[0], "__len__"))

    if has_one_axis(X): X = [X]
    if Y is None:
        X, Y = [[]] * len(X), X
    elif has_one_axis(Y):
        Y = [Y]
    if len(X) != len(Y):
        X = X * len(Y)

    set_figsize(figsize)
    if axes is None:
        axes = d2l.plt.gca()
    axes.cla()
    for x, y, fmt in zip(X, Y, fmts):
        axes.plot(x,y,fmt) if len(x) else axes.plot(y,fmt)
    set_axes(axes, xlabel, ylabel, xlim, ylim, xscale, yscale, legend)
```

Now we can plot the function $u = f(x)$ and its tangent line $y = 2x - 3$ at $x = 1$, where the coefficient 2 is the slope of the tangent line.

```
x = np.arange(0, 3, 0.1)
plot(x, [f(x), 2 * x - 3], 'x', 'f(x)', legend=['f(x)', 'Tangent line (x=1)'])
```

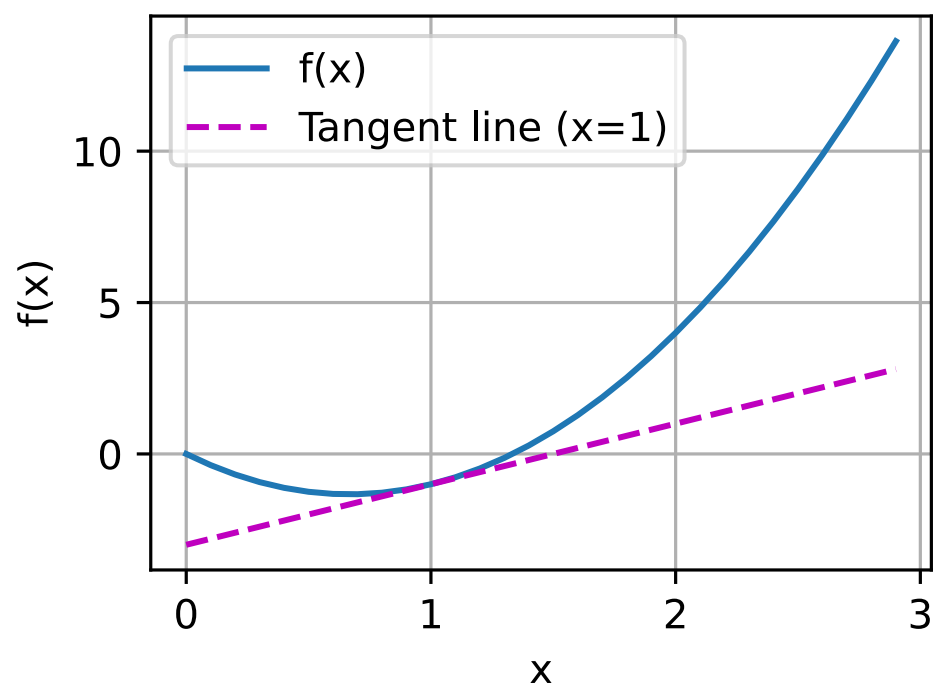

## 2.4.3 Partial Derivatives and Gradients

Thus far, we have been differentiating functions of just one variable. In deep learning, we also need to work with functions of *many* variables. We briefly introduce notions of the derivative that apply to such *multivariate* functions.

Let $y = f(x_1, x_2, \ldots, x_n)$ be a function with $n$ variables. The *partial derivative* of $y$ with respect to its $i^{\text{th}}$ parameter $x_i$ is

$$\frac{\partial y}{\partial x_i} = \lim_{h \to 0} \frac{f(x_1, \ldots, x_{i-1}, x_i + h, x_{i+1}, \ldots, x_n) - f(x_1, \ldots, x_i, \ldots, x_n)}{h}. \tag{2.4.6}$$

To calculate $\frac{\partial y}{\partial x_i}$, we can treat $x_1, \ldots, x_{i-1}, x_{i+1}, \ldots, x_n$ as constants and calculate the derivative of $y$ with respect to $x_i$. The following notational conventions for partial derivatives are all common and all mean the same thing:

$$\frac{\partial y}{\partial x_i} = \frac{\partial f}{\partial x_i} = \partial_{x_i} f = \partial_i f = f_{x_i} = f_i = D_i f = D_{x_i} f. \tag{2.4.7}$$

We can concatenate partial derivatives of a multivariate function with respect to all its variables to obtain a vector that is called the *gradient* of the function. Suppose that the input of function $f : \mathbb{R}^n \to \mathbb{R}$ is an $n$-dimensional vector $\mathbf{x} = [x_1, x_2, \ldots, x_n]^\top$ and the output is a scalar. The gradient of the function $f$ with respect to $\mathbf{x}$ is a vector of $n$ partial derivatives:

$$\nabla_{\mathbf{x}} f(\mathbf{x}) = [\partial_{x_1} f(\mathbf{x}), \partial_{x_2} f(\mathbf{x}), \ldots \partial_{x_n} f(\mathbf{x})]^\top. \tag{2.4.8}$$

When there is no ambiguity, $\nabla_{\mathbf{x}} f(\mathbf{x})$ is typically replaced by $\nabla f(\mathbf{x})$. The following rules come in handy for differentiating multivariate functions:

- For all $\mathbf{A} \in \mathbb{R}^{m \times n}$ we have $\nabla_{\mathbf{x}} \mathbf{A}\mathbf{x} = \mathbf{A}^\top$ and $\nabla_{\mathbf{x}} \mathbf{x}^\top \mathbf{A} = \mathbf{A}$.

- For square matrices $\mathbf{A} \in \mathbb{R}^{n \times n}$ we have that $\nabla_{\mathbf{x}} \mathbf{x}^\top \mathbf{A} \mathbf{x} = (\mathbf{A} + \mathbf{A}^\top)\mathbf{x}$ and in particular $\nabla_{\mathbf{x}} \|\mathbf{x}\|^2 = \nabla_{\mathbf{x}} \mathbf{x}^\top \mathbf{x} = 2\mathbf{x}$.

Similarly, for any matrix $\mathbf{X}$, we have $\nabla_{\mathbf{X}} \|\mathbf{X}\|_{\textrm{F}}^2 = 2\mathbf{X}$.

### 2.4.4 Chain Rule

In deep learning, the gradients of concern are often difficult to calculate because we are working with deeply nested functions (of functions (of functions…)). Fortunately, the *chain rule* takes care of this. Returning to functions of a single variable, suppose that $y = f(g(x))$ and that the underlying functions $y = f(u)$ and $u = g(x)$ are both differentiable. The chain rule states that

$$\frac{dy}{dx} = \frac{dy}{du} \frac{du}{dx}. \tag{2.4.9}$$

Turning back to multivariate functions, suppose that $y = f(\mathbf{u})$ has variables $u_1, u_2, \ldots, u_m$, where each $u_i = g_i(\mathbf{x})$ has variables $x_1, x_2, \ldots, x_n$, i.e., $\mathbf{u} = g(\mathbf{x})$. Then the chain rule states that

$$\frac{\partial y}{\partial x_{i}} = \frac{\partial y}{\partial u_{1}} \frac{\partial u_{1}}{\partial x_{i}} + \frac{\partial y}{\partial u_{2}} \frac{\partial u_{2}}{\partial x_{i}} + \ldots + \frac{\partial y}{\partial u_{m}} \frac{\partial u_{m}}{\partial x_{i}} \ \textrm{ and so } \ \nabla_{\mathbf{x}} y = \mathbf{A} \nabla_{\mathbf{u}} y, \tag{2.4.10}$$

where $\mathbf{A} \in \mathbb{R}^{n \times m}$ is a *matrix* that contains the derivative of vector $\mathbf{u}$ with respect to vector $\mathbf{x}$. Thus, evaluating the gradient requires computing a vector–matrix product. This is one of the key reasons why linear algebra is such an integral building block in building deep learning systems.

### 2.4.5 Discussion

While we have just scratched the surface of a deep topic, a number of concepts already come into focus: first, the composition rules for differentiation can be applied routinely, enabling us to compute gradients *automatically*. This task requires no creativity and thus we can focus our cognitive powers elsewhere. Second, computing the derivatives of vector-valued functions requires us to multiply matrices as we trace the dependency graph of variables from output to input. In particular, this graph is traversed in a *forward* direction when we evaluate a function and in a *backwards* direction when we compute gradients. Later chapters will formally introduce backpropagation, a computational procedure for applying the chain rule.

From the viewpoint of optimization, gradients allow us to determine how to move the parameters of a model in order to lower the loss, and each step of the optimization algorithms used throughout this book will require calculating the gradient.

### 2.4.6 Exercises

1. So far we took the rules for derivatives for granted. Using the definition and limits prove the properties for (i) $f(x) = c$, (ii) $f(x) = x^n$, (iii) $f(x) = e^x$ and (iv) $f(x) = \log x$.

2. In the same vein, prove the product, sum, and quotient rule from first principles.
3. Prove that the constant multiple rule follows as a special case of the product rule.
4. Calculate the derivative of $f(x) = x^x$.
5. What does it mean that $f'(x) = 0$ for some $x$? Give an example of a function $f$ and a location $x$ for which this might hold.
6. Plot the function $y = f(x) = x^3 - \frac{1}{x}$ and plot its tangent line at $x = 1$.
7. Find the gradient of the function $f(\mathbf{x}) = 3x_1^2 + 5e^{x_2}$.
8. What is the gradient of the function $f(\mathbf{x}) = \|\mathbf{x}\|_2$? What happens for $\mathbf{x} = \mathbf{0}$?
9. Can you write out the chain rule for the case where $u = f(x, y, z)$ and $x = x(a, b)$, $y = y(a, b)$, and $z = z(a, b)$?
10. Given a function $f(x)$ that is invertible, compute the derivative of its inverse $f^{-1}(x)$. Here we have that $f^{-1}(f(x)) = x$ and conversely $f(f^{-1}(y)) = y$. Hint: use these properties in your derivation.

56 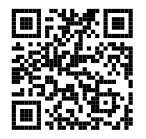 Discussions[56].

## 2.5 Automatic Differentiation

Recall from Section 2.4 that calculating derivatives is the crucial step in all the optimization algorithms that we will use to train deep networks. While the calculations are straightforward, working them out by hand can be tedious and error-prone, and these issues only grow as our models become more complex.

Fortunately all modern deep learning frameworks take this work off our plates by offering *automatic differentiation* (often shortened to *autograd*). As we pass data through each successive function, the framework builds a *computational graph* that tracks how each value depends on others. To calculate derivatives, automatic differentiation works backwards through this graph applying the chain rule. The computational algorithm for applying the chain rule in this fashion is called *backpropagation*.

While autograd libraries have become a hot concern over the past decade, they have a long history. In fact the earliest references to autograd date back over half of a century (Wengert, 1964). The core ideas behind modern backpropagation date to a PhD thesis from 1980 (Speelpenning, 1980) and were further developed in the late 1980s (Griewank, 1989). While backpropagation has become the default method for computing gradients, it is not the only option. For instance, the Julia programming language employs forward propagation (Revels *et al.*, 2016). Before exploring methods, let's first master the autograd package.

```
import torch
```

### 2.5.1 A Simple Function

Let's assume that we are interested in differentiating the function $y = 2\mathbf{x}^\top\mathbf{x}$ with respect to the column vector $\mathbf{x}$. To start, we assign `x` an initial value.

```
x = torch.arange(4.0)
x
```

```
tensor([0., 1., 2., 3.])
```

Before we calculate the gradient of $y$ with respect to $\mathbf{x}$, we need a place to store it. In general, we avoid allocating new memory every time we take a derivative because deep learning requires successively computing derivatives with respect to the same parameters a great many times, and we might risk running out of memory. Note that the gradient of a scalar-valued function with respect to a vector $\mathbf{x}$ is vector-valued with the same shape as $\mathbf{x}$.

```
# Can also create x = torch.arange(4.0, requires_grad=True)
x.requires_grad_(True)
x.grad  # The gradient is None by default
```

We now calculate our function of `x` and assign the result to `y`.

```
y = 2 * torch.dot(x, x)
y
```

```
tensor(28., grad_fn=<MulBackward0>)
```

We can now take the gradient of `y` with respect to `x` by calling its `backward` method. Next, we can access the gradient via `x`'s `grad` attribute.

```
y.backward()
x.grad
```

```
tensor([ 0.,  4.,  8., 12.])
```

We already know that the gradient of the function $y = 2\mathbf{x}^\top\mathbf{x}$ with respect to $\mathbf{x}$ should be $4\mathbf{x}$. We can now verify that the automatic gradient computation and the expected result are identical.

```
x.grad == 4 * x
```

```
tensor([True, True, True, True])
```

Now let's calculate another function of `x` and take its gradient. Note that PyTorch does not automatically reset the gradient buffer when we record a new gradient. Instead, the new gradient is added to the already-stored gradient. This behavior comes in handy when we want to optimize the sum of multiple objective functions. To reset the gradient buffer, we can call `x.grad.zero_()` as follows:

```
x.grad.zero_()  # Reset the gradient
y = x.sum()
y.backward()
x.grad
```

```
tensor([1., 1., 1., 1.])
```

### 2.5.2 Backward for Non-Scalar Variables

When `y` is a vector, the most natural representation of the derivative of `y` with respect to a vector `x` is a matrix called the *Jacobian* that contains the partial derivatives of each component of `y` with respect to each component of `x`. Likewise, for higher-order `y` and `x`, the result of differentiation could be an even higher-order tensor.

While Jacobians do show up in some advanced machine learning techniques, more commonly we want to sum up the gradients of each component of `y` with respect to the full vector `x`, yielding a vector of the same shape as `x`. For example, we often have a vector representing the value of our loss function calculated separately for each example among a *batch* of training examples. Here, we just want to sum up the gradients computed individually for each example.

Because deep learning frameworks vary in how they interpret gradients of non-scalar tensors, PyTorch takes some steps to avoid confusion. Invoking `backward` on a non-scalar elicits an error unless we tell PyTorch how to reduce the object to a scalar. More formally, we need to provide some vector $\mathbf{v}$ such that `backward` will compute $\mathbf{v}^\top \partial_{\mathbf{x}} \mathbf{y}$ rather than $\partial_{\mathbf{x}} \mathbf{y}$. This next part may be confusing, but for reasons that will become clear later, this argument (representing $\mathbf{v}$) is named `gradient`. For a more detailed description, see Yang Zhang's Medium post[57].

```
x.grad.zero_()
y = x * x
y.backward(gradient=torch.ones(len(y)))  # Faster: y.sum().backward()
x.grad
```

```
tensor([0., 2., 4., 6.])
```

### 2.5.3 Detaching Computation

Sometimes, we wish to move some calculations outside of the recorded computational graph. For example, say that we use the input to create some auxiliary intermediate terms for which we do not want to compute a gradient. In this case, we need to *detach* the respective computational graph from the final result. The following toy example makes this clearer: suppose we have `z = x * y` and `y = x * x` but we want to focus on the *direct* influence of `x` on `z` rather than the influence conveyed via `y`. In this case, we can create a new variable `u` that takes the same value as `y` but whose *provenance* (how it was created) has been wiped out. Thus `u` has no ancestors in the graph and gradients do not flow through `u` to `x`. For example, taking the gradient of `z = x * u` will yield the result `u`, (not `3 * x * x` as you might have expected since `z = x * x * x`).

```
x.grad.zero_()
y = x * x
u = y.detach()
z = u * x

z.sum().backward()
x.grad == u
```

```
tensor([True, True, True, True])
```

Note that while this procedure detaches `y`'s ancestors from the graph leading to `z`, the computational graph leading to `y` persists and thus we can calculate the gradient of `y` with respect to `x`.

```
x.grad.zero_()
y.sum().backward()
x.grad == 2 * x
```

```
tensor([True, True, True, True])
```

### 2.5.4 Gradients and Python Control Flow

So far we reviewed cases where the path from input to output was well defined via a function such as `z = x * x * x`. Programming offers us a lot more freedom in how we compute results. For instance, we can make them depend on auxiliary variables or condition choices on intermediate results. One benefit of using automatic differentiation is that even if building the computational graph of a function required passing through a maze of Python control flow (e.g., conditionals, loops, and arbitrary function calls), we can still calculate the gradient of the resulting variable. To illustrate this, consider the following code snippet where the number of iterations of the `while` loop and the evaluation of the `if` statement both depend on the value of the input `a`.

```
def f(a):
    b = a * 2
    while b.norm() < 1000:
        b = b * 2
    if b.sum() > 0:
        c = b
    else:
        c = 100 * b
    return c
```

Below, we call this function, passing in a random value, as input. Since the input is a random variable, we do not know what form the computational graph will take. However, whenever we execute `f(a)` on a specific input, we realize a specific computational graph and can subsequently run `backward`.

```
a = torch.randn(size=(), requires_grad=True)
d = f(a)
d.backward()
```

Even though our function `f` is, for demonstration purposes, a bit contrived, its dependence on the input is quite simple: it is a *linear* function of `a` with piecewise defined scale. As such, `f(a) / a` is a vector of constant entries and, moreover, `f(a) / a` needs to match the gradient of `f(a)` with respect to `a`.

```
a.grad == d / a
```

```
tensor(True)
```

Dynamic control flow is very common in deep learning. For instance, when processing text, the computational graph depends on the length of the input. In these cases, automatic differentiation becomes vital for statistical modeling since it is impossible to compute the gradient *a priori*.

### 2.5.5 Discussion

You have now gotten a taste of the power of automatic differentiation. The development of libraries for calculating derivatives both automatically and efficiently has been a massive productivity booster for deep learning practitioners, liberating them so they can focus on less menial. Moreover, autograd lets us design massive models for which pen and paper gradient computations would be prohibitively time consuming. Interestingly, while we use autograd to *optimize* models (in a statistical sense) the *optimization* of autograd libraries themselves (in a computational sense) is a rich subject of vital interest to framework designers. Here, tools from compilers and graph manipulation are leveraged to compute results in the most expedient and memory-efficient manner.

For now, try to remember these basics: (i) attach gradients to those variables with respect to which we desire derivatives; (ii) record the computation of the target value; (iii) execute the backpropagation function; and (iv) access the resulting gradient.

### 2.5.6 Exercises

1. Why is the second derivative much more expensive to compute than the first derivative?
2. After running the function for backpropagation, immediately run it again and see what happens. Investigate.
3. In the control flow example where we calculate the derivative of `d` with respect to `a`, what would happen if we changed the variable `a` to a random vector or a matrix? At this point, the result of the calculation `f(a)` is no longer a scalar. What happens to the result? How do we analyze this?
4. Let $f(x) = \sin(x)$. Plot the graph of $f$ and of its derivative $f'$. Do not exploit the fact that $f'(x) = \cos(x)$ but rather use automatic differentiation to get the result.
5. Let $f(x) = ((\log x^2) \cdot \sin x) + x^{-1}$. Write out a dependency graph tracing results from $x$ to $f(x)$.
6. Use the chain rule to compute the derivative $\frac{df}{dx}$ of the aforementioned function, placing each term on the dependency graph that you constructed previously.
7. Given the graph and the intermediate derivative results, you have a number of options when computing the gradient. Evaluate the result once starting from $x$ to $f$ and once from $f$ tracing back to $x$. The path from $x$ to $f$ is commonly known as *forward differentiation*, whereas the path from $f$ to $x$ is known as backward differentiation.
8. When might you want to use forward, and when backward, differentiation? Hint: consider the amount of intermediate data needed, the ability to parallelize steps, and the size of matrices and vectors involved.

Discussions[58].

## 2.6 Probability and Statistics

One way or another, machine learning is all about uncertainty. In supervised learning, we want to predict something unknown (the *target*) given something known (the *features*). Depending on our objective, we might attempt to predict the most likely value of the target. Or we might predict the value with the smallest expected distance from the target. And sometimes we wish not only to predict a specific value but to *quantify our uncertainty*. For example, given some features describing a patient, we might want to know *how likely* they are to suffer a heart attack in the next year. In unsupervised learning, we often care about uncertainty. To determine whether a set of measurements are anomalous, it helps to know how likely one is to observe values in a population of interest. Furthermore, in reinforcement learning, we wish to develop agents that act intelligently in various environments. This requires reasoning about how an environment might be expected to change and what rewards one might expect to encounter in response to each of the available actions.

*Probability* is the mathematical field concerned with reasoning under uncertainty. Given a probabilistic model of some process, we can reason about the likelihood of various events. The use of probabilities to describe the frequencies of repeatable events (like coin tosses) is fairly uncontroversial. In fact, *frequentist* scholars adhere to an interpretation of probability that applies *only* to such repeatable events. By contrast *Bayesian* scholars use the language of probability more broadly to formalize reasoning under uncertainty. Bayesian probability is characterized by two unique features: (i) assigning degrees of belief to non-repeatable events, e.g., what is the *probability* that a dam will collapse?; and (ii) subjectivity. While Bayesian probability provides unambiguous rules for how one should update their beliefs in light of new evidence, it allows for different individuals to start off with different *prior* beliefs. *Statistics* helps us to reason backwards, starting off with collection and organization of data and backing out to what inferences we might draw about the process that generated the data. Whenever we analyze a dataset, hunting for patterns that we hope might characterize a broader population, we are employing statistical thinking. Many courses, majors, theses, careers, departments, companies, and institutions have been devoted to the study of probability and statistics. While this section only scratches the surface, we will provide the foundation that you need to begin building models.

```
%matplotlib inline
import random
import torch
from torch.distributions.multinomial import Multinomial
from d2l import torch as d2l
```

## 2.6.1 A Simple Example: Tossing Coins

Imagine that we plan to toss a coin and want to quantify how likely we are to see heads (vs. tails). If the coin is *fair*, then both outcomes (heads and tails), are equally likely. Moreover if we plan to toss the coin $n$ times then the fraction of heads that we *expect* to see should exactly match the *expected* fraction of tails. One intuitive way to see this is by symmetry: for every possible outcome with $n_\textrm{h}$ heads and $n_\textrm{t} = (n - n_\textrm{h})$ tails, there is an equally likely outcome with $n_\textrm{t}$ heads and $n_\textrm{h}$ tails. Note that this is only possible if on average we expect to see $1/2$ of tosses come up heads and $1/2$ come up tails. Of course, if you conduct this experiment many times with $n = 1000000$ tosses each, you might never see a trial where $n_\textrm{h} = n_\textrm{t}$ exactly.

Formally, the quantity $1/2$ is called a *probability* and here it captures the certainty with which any given toss will come up heads. Probabilities assign scores between $0$ and $1$ to outcomes of interest, called *events*. Here the event of interest is heads and we denote the corresponding probability $P(\text{heads})$. A probability of $1$ indicates absolute certainty (imagine a trick coin where both sides were heads) and a probability of $0$ indicates impossibility (e.g., if both sides were tails). The frequencies $n_\text{h}/n$ and $n_\text{t}/n$ are not probabilities but rather *statistics*. Probabilities are *theoretical* quantities that underly the data generating process. Here, the probability $1/2$ is a property of the coin itself. By contrast, statistics are *empirical* quantities that are computed as functions of the observed data. Our interests in probabilistic and statistical quantities are inextricably intertwined. We often design special statistics called *estimators* that, given a dataset, produce *estimates* of model parameters such as probabilities. Moreover, when those estimators satisfy a nice property called *consistency*, our estimates will converge to the corresponding probability. In turn, these inferred probabilities tell about the likely statistical properties of data from the same population that we might encounter in the future.

Suppose that we stumbled upon a real coin for which we did not know the true $P(\text{heads})$. To investigate this quantity with statistical methods, we need to (i) collect some data; and (ii) design an estimator. Data acquisition here is easy; we can toss the coin many times and record all the outcomes. Formally, drawing realizations from some underlying random process is called *sampling*. As you might have guessed, one natural estimator is the ratio of the number of observed *heads* to the total number of tosses.

Now, suppose that the coin was in fact fair, i.e., $P(\text{heads}) = 0.5$. To simulate tosses of a fair coin, we can invoke any random number generator. There are some easy ways to draw samples of an event with probability $0.5$. For example Python's `random.random` yields numbers in the interval $[0, 1]$ where the probability of lying in any sub-interval $[a, b] \subset [0, 1]$ is equal to $b - a$. Thus we can get out `0` and `1` with probability `0.5` each by testing whether the returned float number is greater than `0.5`:

```
num_tosses = 100
heads = sum([random.random() > 0.5 for _ in range(num_tosses)])
tails = num_tosses - heads
print("heads, tails: ", [heads, tails])
```

```
heads, tails:  [45, 55]
```

More generally, we can simulate multiple draws from any variable with a finite number of possible outcomes (like the toss of a coin or roll of a die) by calling the multinomial function, setting the first argument to the number of draws and the second as a list of probabilities associated with each of the possible outcomes. To simulate ten tosses of a fair coin, we assign probability vector `[0.5, 0.5]`, interpreting index 0 as heads and index 1 as tails. The function returns a vector with length equal to the number of possible outcomes (here, 2), where the first component tells us the number of occurrences of heads and the second component tells us the number of occurrences of tails.

```
fair_probs = torch.tensor([0.5, 0.5])
Multinomial(100, fair_probs).sample()
```

```
tensor([44., 56.])
```

Each time you run this sampling process, you will receive a new random value that may differ from the previous outcome. Dividing by the number of tosses gives us the *frequency* of each outcome in our data. Note that these frequencies, just like the probabilities that they are intended to estimate, sum to 1.

```
Multinomial(100, fair_probs).sample() / 100
```

```
tensor([0.4600, 0.5400])
```

Here, even though our simulated coin is fair (we ourselves set the probabilities `[0.5, 0.5]`), the counts of heads and tails may not be identical. That is because we only drew a relatively small number of samples. If we did not implement the simulation ourselves, and only saw the outcome, how would we know if the coin were slightly unfair or if the possible deviation from $1/2$ was just an artifact of the small sample size? Let's see what happens when we simulate 10,000 tosses.

```
counts = Multinomial(10000, fair_probs).sample()
counts / 10000
```

```
tensor([0.4957, 0.5043])
```

In general, for averages of repeated events (like coin tosses), as the number of repetitions grows, our estimates are guaranteed to converge to the true underlying probabilities. The mathematical formulation of this phenomenon is called the *law of large numbers* and the *central limit theorem* tells us that in many situations, as the sample size $n$ grows, these errors should go down at a rate of $(1/\sqrt{n})$. Let's get some more intuition by studying how our estimate evolves as we grow the number of tosses from 1 to 10,000.

```
counts = Multinomial(1, fair_probs).sample((10000,))
cum_counts = counts.cumsum(dim=0)
estimates = cum_counts / cum_counts.sum(dim=1, keepdims=True)
estimates = estimates.numpy()

d2l.set_figsize((4.5, 3.5))
d2l.plt.plot(estimates[:, 0], label=("P(coin=heads)"))
d2l.plt.plot(estimates[:, 1], label=("P(coin=tails)"))
d2l.plt.axhline(y=0.5, color='black', linestyle='dashed')
d2l.plt.gca().set_xlabel('Samples')
d2l.plt.gca().set_ylabel('Estimated probability')
d2l.plt.legend();
```

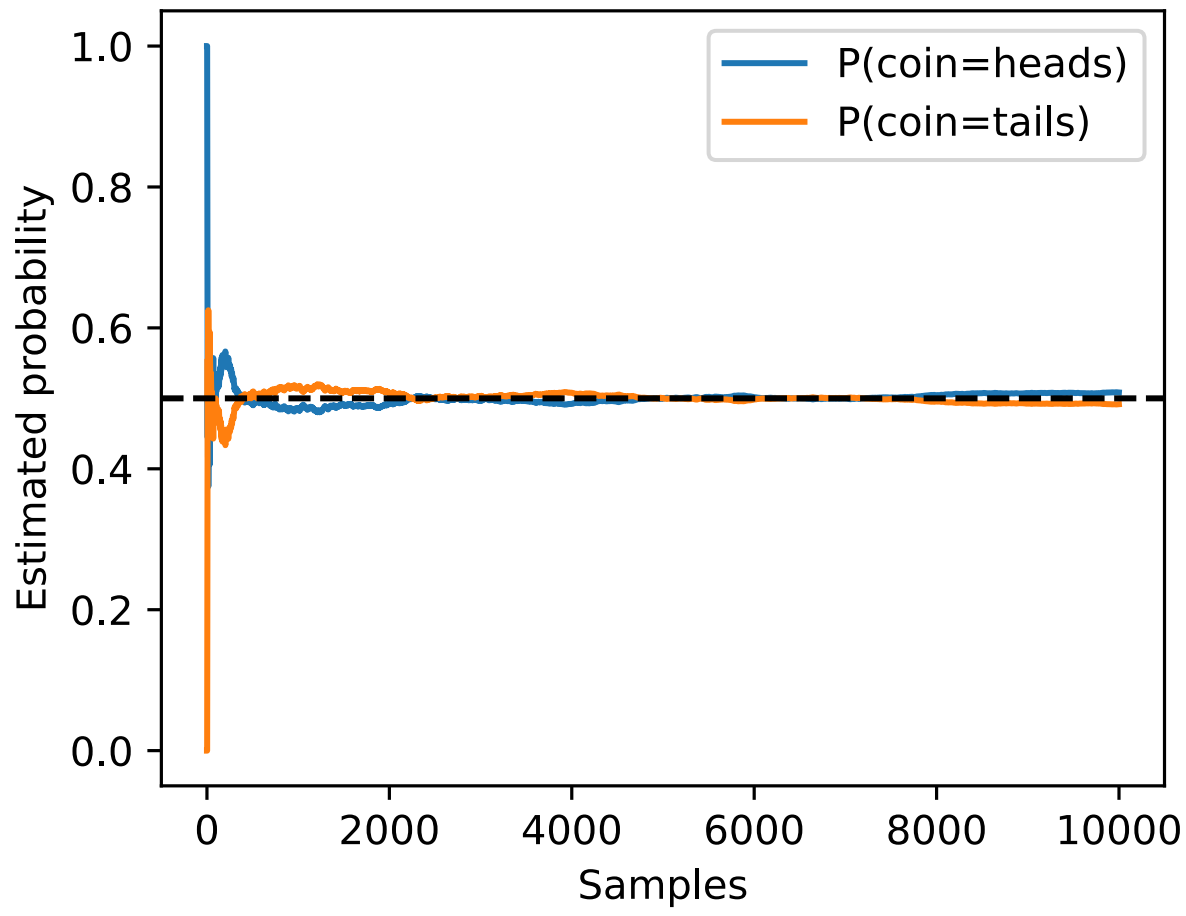

Each solid curve corresponds to one of the two values of the coin and gives our estimated probability that the coin turns up that value after each group of experiments. The dashed black line gives the true underlying probability. As we get more data by conducting more experiments, the curves converge towards the true probability. You might already begin to see the shape of some of the more advanced questions that preoccupy statisticians: How quickly does this convergence happen? If we had already tested many coins manufactured at the same plant, how might we incorporate this information?

### 2.6.2 A More Formal Treatment

We have already gotten pretty far: posing a probabilistic model, generating synthetic data, running a statistical estimator, empirically assessing convergence, and reporting error metrics (checking the deviation). However, to go much further, we will need to be more precise.

When dealing with randomness, we denote the set of possible outcomes $\mathcal{S}$ and call it the *sample space* or *outcome space*. Here, each element is a distinct possible *outcome*. In the case of rolling a single coin, $\mathcal{S} = \{\textrm{heads}, \textrm{tails}\}$. For a single die, $\mathcal{S} = \{1, 2, 3, 4, 5, 6\}$. When flipping two coins, possible outcomes are $\{(\textrm{heads}, \textrm{heads}), (\textrm{heads}, \textrm{tails}), (\textrm{tails}, \textrm{heads}), (\textrm{tails}, \textrm{tails})\}$. *Events* are subsets of the sample space. For instance, the event "the first coin toss comes up heads" corresponds to the set $\{(\textrm{heads}, \textrm{heads}), (\textrm{heads}, \textrm{tails})\}$. Whenever the outcome $z$ of a random experiment satisfies $z \in \mathcal{A}$, then event $\mathcal{A}$ has occurred. For a single roll of a die, we could define the events "seeing a 5" ($\mathcal{A} = \{5\}$) and "seeing an odd number" ($\mathcal{B} = \{1, 3, 5\}$). In this case, if the die came up 5, we would say that both $\mathcal{A}$ and $\mathcal{B}$ occurred. On the other hand, if $z = 3$, then $\mathcal{A}$ did not occur but $\mathcal{B}$ did.

A *probability* function maps events onto real values $P : \mathcal{A} \subseteq \mathcal{S} \rightarrow [0, 1]$. The probability, denoted $P(\mathcal{A})$, of an event $\mathcal{A}$ in the given sample space $\mathcal{S}$, has the following properties:

- The probability of any event $\mathcal{A}$ is a nonnegative real number, i.e., $P(\mathcal{A}) \geq 0$;

- The probability of the entire sample space is 1, i.e., $P(\mathcal{S}) = 1$;
- For any countable sequence of events $\mathcal{A}_1, \mathcal{A}_2, \ldots$ that are *mutually exclusive* (i.e., $\mathcal{A}_i \cap \mathcal{A}_j = \emptyset$ for all $i \neq j$), the probability that any of them happens is equal to the sum of their individual probabilities, i.e., $P(\bigcup_{i=1}^{\infty} \mathcal{A}_i) = \sum_{i=1}^{\infty} P(\mathcal{A}_i)$.

These axioms of probability theory, proposed by Kolmogorov (1933), can be applied to rapidly derive a number of important consequences. For instance, it follows immediately that the probability of any event $\mathcal{A}$ *or* its complement $\mathcal{A}'$ occurring is 1 (because $\mathcal{A} \cup \mathcal{A}' = \mathcal{S}$). We can also prove that $P(\emptyset) = 0$ because $1 = P(\mathcal{S} \cup \mathcal{S}') = P(\mathcal{S} \cup \emptyset) = P(\mathcal{S}) + P(\emptyset) = 1 + P(\emptyset)$. Consequently, the probability of any event $\mathcal{A}$ *and* its complement $\mathcal{A}'$ occurring simultaneously is $P(\mathcal{A} \cap \mathcal{A}') = 0$. Informally, this tells us that impossible events have zero probability of occurring.

### 2.6.3 Random Variables

When we spoke about events like the roll of a die coming up odds or the first coin toss coming up heads, we were invoking the idea of a *random variable*. Formally, random variables are mappings from an underlying sample space to a set of (possibly many) values. You might wonder how a random variable is different from the sample space, since both are collections of outcomes. Importantly, random variables can be much coarser than the raw sample space. We can define a binary random variable like “greater than 0.5” even when the underlying sample space is infinite, e.g., points on the line segment between 0 and 1. Additionally, multiple random variables can share the same underlying sample space. For example “whether my home alarm goes off” and “whether my house was burgled” are both binary random variables that share an underlying sample space. Consequently, knowing the value taken by one random variable can tell us something about the likely value of another random variable. Knowing that the alarm went off, we might suspect that the house was likely burgled.

Every value taken by a random variable corresponds to a subset of the underlying sample space. Thus the occurrence where the random variable $X$ takes value $v$, denoted by $X = v$, is an *event* and $P(X = v)$ denotes its probability. Sometimes this notation can get clunky, and we can abuse notation when the context is clear. For example, we might use $P(X)$ to refer broadly to the *distribution* of $X$, i.e., the function that tells us the probability that $X$ takes any given value. Other times we write expressions like $P(X, Y) = P(X)P(Y)$, as a shorthand to express a statement that is true for all of the values that the random variables $X$ and $Y$ can take, i.e., for all $i, j$ it holds that $P(X = i \text{ and } Y = j) = P(X = i)P(Y = j)$. Other times, we abuse notation by writing $P(v)$ when the random variable is clear from the context. Since an event in probability theory is a set of outcomes from the sample space, we can specify a range of values for a random variable to take. For example, $P(1 \leq X \leq 3)$ denotes the probability of the event $\{1 \leq X \leq 3\}$.

Note that there is a subtle difference between *discrete* random variables, like flips of a coin or tosses of a die, and *continuous* ones, like the weight and the height of a person sampled at random from the population. In this case we seldom really care about someone’s exact height. Moreover, if we took precise enough measurements, we would find that no two people on the planet have the exact same height. In fact, with fine enough measurements, you would never have the same height when you wake up and when you go to sleep. There is little point in asking about the exact probability that someone is 1.801392782910287192 meters tall. Instead, we typically care more about being able to say whether someone's height falls into a given interval, say between 1.79 and 1.81 meters. In these cases we work with probability *densities*. The height of exactly 1.80 meters has no probability, but nonzero density. To work out the probability assigned to an interval, we must take an *integral* of the density over that interval.

### 2.6.4 Multiple Random Variables

You might have noticed that we could not even make it through the previous section without making statements involving interactions among multiple random variables (recall $P(X, Y) = P(X)P(Y)$). Most of machine learning is concerned with such relationships. Here, the sample space would be the population of interest, say customers who transact with a business, photographs on the Internet, or proteins known to biologists. Each random variable would represent the (unknown) value of a different attribute. Whenever we sample an individual from the population, we observe a realization of each of the random variables. Because the values taken by random variables correspond to subsets of the sample space that could be overlapping, partially overlapping, or entirely disjoint, knowing the value taken by one random variable can cause us to update our beliefs about which values of another random variable are likely. If a patient walks into a hospital and we observe that they are having trouble breathing and have lost their sense of smell, then we believe that they are more likely to have COVID-19 than we might if they had no trouble breathing and a perfectly ordinary sense of smell.

When working with multiple random variables, we can construct events corresponding to every combination of values that the variables can jointly take. The probability function that assigns probabilities to each of these combinations (e.g. $A = a$ and $B = b$) is called the *joint probability* function and simply returns the probability assigned to the intersection of the corresponding subsets of the sample space. The *joint probability* assigned to the event where random variables $A$ and $B$ take values $a$ and $b$, respectively, is denoted $P(A = a, B = b)$, where the comma indicates "and". Note that for any values $a$ and $b$, it follows that

$$P(A = a, B = b) \leq P(A = a) \text{ and } P(A = a, B = b) \leq P(B = b), \tag{2.6.1}$$

since for $A = a$ and $B = b$ to happen, $A = a$ has to happen *and* $B = b$ also has to happen. Interestingly, the joint probability tells us all that we can know about these random variables in a probabilistic sense, and can be used to derive many other useful quantities, including recovering the individual distributions $P(A)$ and $P(B)$. To recover $P(A = a)$ we simply sum up $P(A = a, B = v)$ over all values $v$ that the random variable $B$ can take: $P(A = a) = \sum_v P(A = a, B = v)$.

The ratio $\frac{P(A=a,B=b)}{P(A=a)} \leq 1$ turns out to be extremely important. It is called the *conditional*

*probability*, and is denoted via the "|" symbol:

$$P(B = b \mid A = a) = P(A = a, B = b)/P(A = a). \tag{2.6.2}$$

It tells us the new probability associated with the event $B = b$, once we condition on the fact $A = a$ took place. We can think of this conditional probability as restricting attention only to the subset of the sample space associated with $A = a$ and then renormalizing so that all probabilities sum to 1. Conditional probabilities are in fact just ordinary probabilities and thus respect all of the axioms, as long as we condition all terms on the same event and thus restrict attention to the same sample space. For instance, for disjoint events $\mathcal{B}$ and $\mathcal{B}'$, we have that $P(\mathcal{B} \cup \mathcal{B}' \mid A = a) = P(\mathcal{B} \mid A = a) + P(\mathcal{B}' \mid A = a)$.

Using the definition of conditional probabilities, we can derive the famous result called *Bayes' theorem*. By construction, we have that $P(A, B) = P(B \mid A)P(A)$ and $P(A, B) = P(A \mid B)P(B)$. Combining both equations yields $P(B \mid A)P(A) = P(A \mid B)P(B)$ and hence

$$P(A \mid B) = \frac{P(B \mid A)P(A)}{P(B)}. \tag{2.6.3}$$

This simple equation has profound implications because it allows us to reverse the order of conditioning. If we know how to estimate $P(B \mid A)$, $P(A)$, and $P(B)$, then we can estimate $P(A \mid B)$. We often find it easier to estimate one term directly but not the other and Bayes' theorem can come to the rescue here. For instance, if we know the prevalence of symptoms for a given disease, and the overall prevalences of the disease and symptoms, respectively, we can determine how likely someone is to have the disease based on their symptoms. In some cases we might not have direct access to $P(B)$, such as the prevalence of symptoms. In this case a simplified version of Bayes' theorem comes in handy:

$$P(A \mid B) \propto P(B \mid A)P(A). \tag{2.6.4}$$

Since we know that $P(A \mid B)$ must be normalized to 1, i.e., $\sum_a P(A = a \mid B) = 1$, we can use it to compute

$$P(A \mid B) = \frac{P(B \mid A)P(A)}{\sum_a P(B \mid A = a)P(A = a)}. \tag{2.6.5}$$

In Bayesian statistics, we think of an observer as possessing some (subjective) prior beliefs about the plausibility of the available hypotheses encoded in the *prior* $P(H)$, and a *likelihood function* that says how likely one is to observe any value of the collected evidence for each of the hypotheses in the class $P(E \mid H)$. Bayes' theorem is then interpreted as telling us how to update the initial *prior* $P(H)$ in light of the available evidence $E$ to produce *posterior* beliefs $P(H \mid E) = \frac{P(E|H)P(H)}{P(E)}$. Informally, this can be stated as "posterior equals prior times likelihood, divided by the evidence". Now, because the evidence $P(E)$ is the same for all hypotheses, we can get away with simply normalizing over the hypotheses.

Note that $\sum_a P(A = a \mid B) = 1$ also allows us to *marginalize* over random variables. That is, we can drop variables from a joint distribution such as $P(A, B)$. After all, we have that

$$\sum_a P(B \mid A = a)P(A = a) = \sum_a P(B, A = a) = P(B). \tag{2.6.6}$$

Independence is another fundamentally important concept that forms the backbone of many important ideas in statistics. In short, two variables are *independent* if conditioning on the value of $A$ does not cause any change to the probability distribution associated with $B$ and vice versa. More formally, independence, denoted $A \perp B$, requires that $P(A \mid B) = P(A)$ and, consequently, that $P(A, B) = P(A \mid B)P(B) = P(A)P(B)$. Independence is often an appropriate assumption. For example, if the random variable $A$ represents the outcome from tossing one fair coin and the random variable $B$ represents the outcome from tossing another, then knowing whether $A$ came up heads should not influence the probability of $B$ coming up heads.

Independence is especially useful when it holds among the successive draws of our data from some underlying distribution (allowing us to make strong statistical conclusions) or when it holds among various variables in our data, allowing us to work with simpler models that encode this independence structure. On the other hand, estimating the dependencies among random variables is often the very aim of learning. We care to estimate the probability of disease given symptoms specifically because we believe that diseases and symptoms are *not* independent.

Note that because conditional probabilities are proper probabilities, the concepts of independence and dependence also apply to them. Two random variables $A$ and $B$ are *conditionally independent* given a third variable $C$ if and only if $P(A, B \mid C) = P(A \mid C)P(B \mid C)$. Interestingly, two variables can be independent in general but become dependent when conditioning on a third. This often occurs when the two random variables $A$ and $B$ correspond to causes of some third variable $C$. For example, broken bones and lung cancer might be independent in the general population but if we condition on being in the hospital then we might find that broken bones are negatively correlated with lung cancer. That is because the broken bone *explains away* why some person is in the hospital and thus lowers the probability that they are hospitalized because of having lung cancer.

And conversely, two dependent random variables can become independent upon conditioning on a third. This often happens when two otherwise unrelated events have a common cause. Shoe size and reading level are highly correlated among elementary school students, but this correlation disappears if we condition on age.

### 2.6.5 An Example

Let's put our skills to the test. Assume that a doctor administers an HIV test to a patient. This test is fairly accurate and fails only with 1% probability if the patient is healthy but reported as diseased, i.e., healthy patients test positive in 1% of cases. Moreover, it never fails to detect HIV if the patient actually has it. We use $D_1 \in \{0, 1\}$ to indicate the diagnosis (0 if negative and 1 if positive) and $H \in \{0, 1\}$ to denote the HIV status.

| Conditional probability | $H = 1$ | $H = 0$ |
|---|---|---|
| $P(D_1 = 1 \mid H)$ | 1 | 0.01 |
| $P(D_1 = 0 \mid H)$ | 0 | 0.99 |

Note that the column sums are all 1 (but the row sums do not), since they are conditional probabilities. Let's compute the probability of the patient having HIV if the test comes back positive, i.e., $P(H = 1 \mid D_1 = 1)$. Intuitively this is going to depend on how common the disease is, since it affects the number of false alarms. Assume that the population is fairly free of the disease, e.g., $P(H = 1) = 0.0015$. To apply Bayes' theorem, we need to apply marginalization to determine

$$\begin{aligned}P(D_1 = 1) =& P(D_1 = 1, H = 0) + P(D_1 = 1, H = 1)\\ =& P(D_1 = 1 \mid H = 0)P(H = 0) + P(D_1 = 1 \mid H = 1)P(H = 1)\\ =& 0.011485.\end{aligned} \tag{2.6.7}$$

This leads us to

$$P(H = 1 \mid D_1 = 1) = \frac{P(D_1 = 1 \mid H = 1)P(H = 1)}{P(D_1 = 1)} = 0.1306. \tag{2.6.8}$$

In other words, there is only a 13.06% chance that the patient actually has HIV, despite the test being pretty accurate. As we can see, probability can be counterintuitive. What should a patient do upon receiving such terrifying news? Likely, the patient would ask the physician to administer another test to get clarity. The second test has different characteristics and it is not as good as the first one.

| Conditional probability | $H = 1$ | $H = 0$ |
|---|---|---|
| $P(D_2 = 1 \mid H)$ | 0.98 | 0.03 |
| $P(D_2 = 0 \mid H)$ | 0.02 | 0.97 |

Unfortunately, the second test comes back positive, too. Let's calculate the requisite probabilities to invoke Bayes' theorem by assuming conditional independence:

$$\begin{aligned}P(D_1 = 1, D_2 = 1 \mid H = 0) = P(D_1 = 1 \mid H = 0)P(D_2 = 1 \mid H = 0) =& \quad 0.0003,\\ P(D_1 = 1, D_2 = 1 \mid H = 1) = P(D_1 = 1 \mid H = 1)P(D_2 = 1 \mid H = 1) =& \quad 0.98.\end{aligned} \tag{2.6.9}$$

Now we can apply marginalization to obtain the probability that both tests come back positive:

$$\begin{aligned}&P(D_1 = 1, D_2 = 1)\\ &= P(D_1 = 1, D_2 = 1, H = 0) + P(D_1 = 1, D_2 = 1, H = 1)\\ &= P(D_1 = 1, D_2 = 1 \mid H = 0)P(H = 0) + P(D_1 = 1, D_2 = 1 \mid H = 1)P(H = 1)\\ &= 0.00176955.\end{aligned} \tag{2.6.10}$$

Finally, the probability of the patient having HIV given that both tests are positive is

$$P(H = 1 \mid D_1 = 1, D_2 = 1) = \frac{P(D_1 = 1, D_2 = 1 \mid H = 1)P(H = 1)}{P(D_1 = 1, D_2 = 1)} = 0.8307. \tag{2.6.11}$$

That is, the second test allowed us to gain much higher confidence that not all is well. Despite the second test being considerably less accurate than the first one, it still significantly improved our estimate. The assumption of both tests being conditionally independent of each other was crucial for our ability to generate a more accurate estimate. Take the extreme case where we run the same test twice. In this situation we would expect the same outcome both times, hence no additional insight is gained from running the same test again. The astute reader might have noticed that the diagnosis behaved like a classifier hiding in plain sight where our ability to decide whether a patient is healthy increases as we obtain more features (test outcomes).

### 2.6.6 Expectations

Often, making decisions requires not just looking at the probabilities assigned to individual events but composing them together into useful aggregates that can provide us with guidance. For example, when random variables take continuous scalar values, we often care about knowing what value to expect *on average*. This quantity is formally called an *expectation*. If we are making investments, the first quantity of interest might be the return we can expect, averaging over all the possible outcomes (and weighting by the appropriate probabilities). For instance, say that with 50% probability, an investment might fail altogether, with 40% probability it might provide a 2× return, and with 10% probability it might provide a 10× return 10×. To calculate the expected return, we sum over all returns, multiplying each by the probability that they will occur. This yields the expectation $0.5 \cdot 0 + 0.4 \cdot 2 + 0.1 \cdot 10 = 1.8$. Hence the expected return is $1.8\times$.

In general, the *expectation* (or average) of the random variable $X$ is defined as

$$E[X] = E_{x \sim P}[x] = \sum_x x P(X = x). \tag{2.6.12}$$

Likewise, for densities we obtain $E[X] = \int x \; dp(x)$. Sometimes we are interested in the expected value of some function of $x$. We can calculate these expectations as

$$E_{x \sim P}[f(x)] = \sum_x f(x)P(x) \text{ and } E_{x \sim P}[f(x)] = \int f(x)p(x) \; dx \tag{2.6.13}$$

for discrete probabilities and densities, respectively. Returning to the investment example from above, $f$ might be the *utility* (happiness) associated with the return. Behavior economists have long noted that people associate greater disutility with losing money than the utility gained from earning one dollar relative to their baseline. Moreover, the value of money tends to be sub-linear. Possessing 100k dollars versus zero dollars can make the difference between paying the rent, eating well, and enjoying quality healthcare versus suffering through homelessness. On the other hand, the gains due to possessing 200k versus 100k are less dramatic. Reasoning like this motivates the cliché that "the utility of money is logarithmic".

If the utility associated with a total loss were $-1$, and the utilities associated with returns of $1$, $2$, and $10$ were $1$, $2$ and $4$, respectively, then the expected happiness of investing would be $0.5 \cdot (-1) + 0.4 \cdot 2 + 0.1 \cdot 4 = 0.7$ (an expected loss of utility of 30%). If indeed this were your utility function, you might be best off keeping the money in the bank.

For financial decisions, we might also want to measure how *risky* an investment is. Here, we care not just about the expected value but how much the actual values tend to *vary* relative to this value. Note that we cannot just take the expectation of the difference between the actual and expected values. This is because the expectation of a difference is the difference of the expectations, i.e., $E[X - E[X]] = E[X] - E[E[X]] = 0$. However, we can look at the expectation of any non-negative function of this difference. The *variance* of a random variable is calculated by looking at the expected value of the *squared* differences:

$$\mathrm{Var}[X] = E\left[(X - E[X])^2\right] = E[X^2] - E[X]^2. \tag{2.6.14}$$

Here the equality follows by expanding $(X - E[X])^2 = X^2 - 2XE[X] + E[X]^2$ and taking expectations for each term. The square root of the variance is another useful quantity called the *standard deviation*. While this and the variance convey the same information (either can be calculated from the other), the standard deviation has the nice property that it is expressed in the same units as the original quantity represented by the random variable.

Lastly, the variance of a function of a random variable is defined analogously as

$$\mathrm{Var}_{x \sim P}[f(x)] = E_{x \sim P}[f^2(x)] - E_{x \sim P}[f(x)]^2. \tag{2.6.15}$$

Returning to our investment example, we can now compute the variance of the investment. It is given by $0.5 \cdot 0 + 0.4 \cdot 2^2 + 0.1 \cdot 10^2 - 1.8^2 = 8.36$. For all intents and purposes this is a risky investment. Note that by mathematical convention mean and variance are often referenced as $\mu$ and $\sigma^2$. This is particularly the case whenever we use it to parametrize a Gaussian distribution.

In the same way as we introduced expectations and variance for *scalar* random variables, we can do so for vector-valued ones. Expectations are easy, since we can apply them elementwise. For instance, $\boldsymbol{\mu} \stackrel{\mathrm{def}}{=} E_{\mathbf{x} \sim P}[\mathbf{x}]$ has coordinates $\mu_i = E_{\mathbf{x} \sim P}[x_i]$. *Covariances* are more complicated. We define them by taking expectations of the *outer product* of the difference between random variables and their mean:

$$\boldsymbol{\Sigma} \stackrel{\mathrm{def}}{=} \mathrm{Cov}_{\mathbf{x} \sim P}[\mathbf{x}] = E_{\mathbf{x} \sim P}\left[(\mathbf{x} - \boldsymbol{\mu})(\mathbf{x} - \boldsymbol{\mu})^\top\right]. \tag{2.6.16}$$

This matrix $\boldsymbol{\Sigma}$ is referred to as the covariance matrix. An easy way to see its effect is to consider some vector $\mathbf{v}$ of the same size as $\mathbf{x}$. It follows that

$$\mathbf{v}^\top \boldsymbol{\Sigma} \mathbf{v} = E_{\mathbf{x} \sim P}\left[\mathbf{v}^\top (\mathbf{x} - \boldsymbol{\mu})(\mathbf{x} - \boldsymbol{\mu})^\top \mathbf{v}\right] = \mathrm{Var}_{x \sim P}[\mathbf{v}^\top \mathbf{x}]. \tag{2.6.17}$$

As such, $\boldsymbol{\Sigma}$ allows us to compute the variance for any linear function of $\mathbf{x}$ by a simple matrix multiplication. The off-diagonal elements tell us how correlated the coordinates are: a value of 0 means no correlation, where a larger positive value means that they are more strongly correlated.

### 2.6.7 Discussion

In machine learning, there are many things to be uncertain about! We can be uncertain about the value of a label given an input. We can be uncertain about the estimated value of a pa-rameter. We can even be uncertain about whether data arriving at deployment is even from the same distribution as the training data.

By *aleatoric uncertainty*, we mean uncertainty that is intrinsic to the problem, and due to genuine randomness unaccounted for by the observed variables. By *epistemic uncertainty*, we mean uncertainty over a model's parameters, the sort of uncertainty that we can hope to reduce by collecting more data. We might have epistemic uncertainty concerning the probability that a coin turns up heads, but even once we know this probability, we are left with aleatoric uncertainty about the outcome of any future toss. No matter how long we watch someone tossing a fair coin, we will never be more or less than 50% certain that the next toss will come up heads. These terms come from mechanical modeling, (see e.g., Der Kiureghian and Ditlevsen (2009) for a review on this aspect of uncertainty quantification[59]). It is worth noting, however, that these terms constitute a slight abuse of language. The term *epistemic* refers to anything concerning *knowledge* and thus, in the philosophical sense, all uncertainty is epistemic.

We saw that sampling data from some unknown probability distribution can provide us with information that can be used to estimate the parameters of the data generating distribution. That said, the rate at which this is possible can be quite slow. In our coin tossing example (and many others) we can do no better than to design estimators that converge at a rate of $1/\sqrt{n}$, where $n$ is the sample size (e.g., the number of tosses). This means that by going from 10 to 1000 observations (usually a very achievable task) we see a tenfold reduction of uncertainty, whereas the next 1000 observations help comparatively little, offering only a 1.41 times reduction. This is a persistent feature of machine learning: while there are often easy gains, it takes a very large amount of data, and often with it an enormous amount of computation, to make further gains. For an empirical review of this fact for large scale language models see Revels *et al.* (2016).

We also sharpened our language and tools for statistical modeling. In the process of that we learned about conditional probabilities and about one of the most important equations in statistics—Bayes' theorem. It is an effective tool for decoupling information conveyed by data through a likelihood term $P(B \mid A)$ that addresses how well observations $B$ match a choice of parameters $A$, and a prior probability $P(A)$ which governs how plausible a particular choice of $A$ was in the first place. In particular, we saw how this rule can be applied to assign probabilities to diagnoses, based on the efficacy of the test *and* the prevalence of the disease itself (i.e., our prior).

Lastly, we introduced a first set of nontrivial questions about the effect of a specific probability distribution, namely expectations and variances. While there are many more than just linear and quadratic expectations for a probability distribution, these two already provide a good deal of knowledge about the possible behavior of the distribution. For instance, Chebyshev's inequality[60] states that $P(|X - \mu| \geq k\sigma) \leq 1/k^2$, where $\mu$ is the expectation, $\sigma^2$ is the variance of the distribution, and $k > 1$ is a confidence parameter of our choosing. It tells us that draws from a distribution lie with at least 50% probability within a $[-\sqrt{2}\sigma, \sqrt{2}\sigma]$ interval centered on the expectation.

60

### 2.6.8 Exercises

1. Give an example where observing more data can reduce the amount of uncertainty about the outcome to an arbitrarily low level.

2. Give an example where observing more data will only reduce the amount of uncertainty up to a point and then no further. Explain why this is the case and where you expect this point to occur.

3. We empirically demonstrated convergence to the mean for the toss of a coin. Calculate the variance of the estimate of the probability that we see a head after drawing $n$ samples.

    1. How does the variance scale with the number of observations?

    2. Use Chebyshev's inequality to bound the deviation from the expectation.

    3. How does it relate to the central limit theorem?

4. Assume that we draw $m$ samples $x_i$ from a probability distribution with zero mean and unit variance. Compute the averages $z_m \stackrel{\text{def}}{=} m^{-1} \sum_{i=1}^{m} x_i$. Can we apply Chebyshev's inequality for every $z_m$ independently? Why not?

61 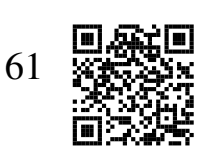

5. Given two events with probability $P(\mathcal{A})$ and $P(\mathcal{B})$, compute upper and lower bounds on $P(\mathcal{A} \cup \mathcal{B})$ and $P(\mathcal{A} \cap \mathcal{B})$. Hint: graph the situation using a Venn diagram[61].

62

6. Assume that we have a sequence of random variables, say $A$, $B$, and $C$, where $B$ only depends on $A$, and $C$ only depends on $B$, can you simplify the joint probability $P(A, B, C)$? Hint: this is a Markov chain[62].

7. In Section 2.6.5, assume that the outcomes of the two tests are not independent. In particular assume that either test on its own has a false positive rate of 10% and a false negative rate of 1%. That is, assume that $P(D = 1 \mid H = 0) = 0.1$ and that $P(D = 0 \mid H = 1) = 0.01$. Moreover, assume that for $H = 1$ (infected) the test outcomes are conditionally independent, i.e., that $P(D_1, D_2 \mid H = 1) = P(D_1 \mid H = 1)P(D_2 \mid H = 1)$ but that for healthy patients the outcomes are coupled via $P(D_1 = D_2 = 1 \mid H = 0) = 0.02$.

    1. Work out the joint probability table for $D_1$ and $D_2$, given $H = 0$ based on the information you have so far.

    2. Derive the probability that the patient is diseased ($H = 1$) after one test returns positive. You can assume the same baseline probability $P(H = 1) = 0.0015$ as before.

    3. Derive the probability that the patient is diseased ($H = 1$) after both tests return positive.

8. Assume that you are an asset manager for an investment bank and you have a choice of stocks $s_i$ to invest in. Your portfolio needs to add up to 1 with weights $\alpha_i$ for each stock. The stocks have an average return $\boldsymbol{\mu} = E_{\mathbf{s} \sim P}[\mathbf{s}]$ and covariance $\boldsymbol{\Sigma} = \mathrm{Cov}_{\mathbf{s} \sim P}[\mathbf{s}]$.

    1. Compute the expected return for a given portfolio $\boldsymbol{\alpha}$.

2. If you wanted to maximize the return of the portfolio, how should you choose your investment?

3. Compute the *variance* of the portfolio.

4. Formulate an optimization problem of maximizing the return while keeping the variance constrained to an upper bound. This is the Nobel-Prize winning Markovitz portfolio[63] (Mangram, 2013). To solve it you will need a quadratic programming solver, something way beyond the scope of this book.

63 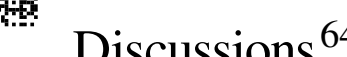

Discussions[64].

64

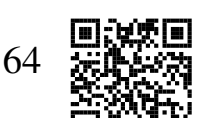

## 2.7 Documentation

While we cannot possibly introduce every single PyTorch function and class (and the information might become outdated quickly), the API documentation[65] and additional tutorials [66] and examples provide such documentation. This section provides some guidance for how to explore the PyTorch API.

```
import torch
```

66

### 2.7.1 Functions and Classes in a Module

To know which functions and classes can be called in a module, we invoke the `dir` function. For instance, we can query all properties in the module for generating random numbers:

```
print(dir(torch.distributions))
```

```
['AbsTransform', 'AffineTransform', 'Bernoulli', 'Beta', 'Binomial',
↪'CatTransform', 'Categorical', 'Cauchy', 'Chi2', 'ComposeTransform',
↪'ContinuousBernoulli', 'CorrCholeskyTransform',
↪'CumulativeDistributionTransform', 'Dirichlet', 'Distribution', 'ExpTransform
↪', 'Exponential', 'ExponentialFamily', 'FisherSnedecor', 'Gamma', 'Geometric
↪', 'Gumbel', 'HalfCauchy', 'HalfNormal', 'Independent', 'IndependentTransform
↪', 'Kumaraswamy', 'LKJCholesky', 'Laplace', 'LogNormal', 'LogisticNormal',
↪'LowRankMultivariateNormal', 'LowerCholeskyTransform', 'MixtureSameFamily',
↪'Multinomial', 'MultivariateNormal', 'NegativeBinomial', 'Normal',
↪'OneHotCategorical', 'OneHotCategoricalStraightThrough', 'Pareto', 'Poisson',
↪ 'PositiveDefiniteTransform', 'PowerTransform', 'RelaxedBernoulli',
↪'RelaxedOneHotCategorical', 'ReshapeTransform', 'SigmoidTransform',
↪'SoftmaxTransform', 'SoftplusTransform', 'StackTransform',
↪'StickBreakingTransform', 'StudentT', 'TanhTransform', 'Transform',
```

```
↪'TransformedDistribution', 'Uniform', 'VonMises', 'Weibull', 'Wishart', '__
↪all__', '__builtins__', '__cached__', '__doc__', '__file__', '__loader__', '_
↪_name__', '__package__', '__path__', '__spec__', 'bernoulli', 'beta',
↪'biject_to', 'binomial', 'categorical', 'cauchy', 'chi2', 'constraint_
↪registry', 'constraints', 'continuous_bernoulli', 'dirichlet', 'distribution
↪', 'exp_family', 'exponential', 'fishersnedecor', 'gamma', 'geometric',
↪'gumbel', 'half_cauchy', 'half_normal', 'identity_transform', 'independent',
↪'kl', 'kl_divergence', 'kumaraswamy', 'laplace', 'lkj_cholesky', 'log_normal
↪', 'logistic_normal', 'lowrank_multivariate_normal', 'mixture_same_family',
↪'multinomial', 'multivariate_normal', 'negative_binomial', 'normal', 'one_
↪hot_categorical', 'pareto', 'poisson', 'register_kl', 'relaxed_bernoulli',
↪'relaxed_categorical', 'studentT', 'transform_to', 'transformed_distribution
↪', 'transforms', 'uniform', 'utils', 'von_mises', 'weibull', 'wishart']
```

Generally, we can ignore functions that start and end with `__` (special objects in Python) or functions that start with a single `_`(usually internal functions). Based on the remaining function or attribute names, we might hazard a guess that this module offers various methods for generating random numbers, including sampling from the uniform distribution (`uniform`), normal distribution (`normal`), and multinomial distribution (`multinomial`).

### 2.7.2 Specific Functions and Classes

For specific instructions on how to use a given function or class, we can invoke the `help` function. As an example, let's explore the usage instructions for tensors' `ones` function.

```
help(torch.ones)
```

```
Help on built-in function ones in module torch:

ones(...)
    ones(*size, *, out=None, dtype=None, layout=torch.strided, device=None,
↪ requires_grad=False) -> Tensor

    Returns a tensor filled with the scalar value 1, with the shape defined
    by the variable argument size.

    Args:
        size (int...): a sequence of integers defining the shape of the␣
↪output tensor.
            Can be a variable number of arguments or a collection like a␣
↪list or tuple.

    Keyword arguments:
        out (Tensor, optional): the output tensor.
        dtype (torch.dtype, optional): the desired data type of returned␣
↪tensor.
```

```
            Default: if None, uses a global default (see torch.set_default_
↪tensor_type()).
        layout (torch.layout, optional): the desired layout of returned␣
↪Tensor.
            Default: torch.strided.
        device (torch.device, optional): the desired device of returned␣
↪tensor.
            Default: if None, uses the current device for the default tensor␣
↪type
            (see torch.set_default_tensor_type()). device will be the CPU
            for CPU tensor types and the current CUDA device for CUDA tensor␣
↪types.
        requires_grad (bool, optional): If autograd should record operations␣
↪on the
            returned tensor. Default: False.

    Example::

        >>> torch.ones(2, 3)
        tensor([[ 1.,  1.,  1.],
                [ 1.,  1.,  1.]])

        >>> torch.ones(5)
        tensor([ 1.,  1.,  1.,  1.,  1.])
```

From the documentation, we can see that the `ones` function creates a new tensor with the specified shape and sets all the elements to the value of 1. Whenever possible, you should run a quick test to confirm your interpretation:

```
torch.ones(4)
```

```
tensor([1., 1., 1., 1.])
```

In the Jupyter notebook, we can use `?` to display the document in another window. For example, `list?` will create content that is almost identical to `help(list)`, displaying it in a new browser window. In addition, if we use two question marks, such as `list??`, the Python code implementing the function will also be displayed.

The official documentation provides plenty of descriptions and examples that are beyond this book. We emphasize important use cases that will get you started quickly with practical problems, rather than completeness of coverage. We also encourage you to study the source code of the libraries to see examples of high-quality implementations of production code. By doing this you will become a better engineer in addition to becoming a better scientist.

Discussions[67].

67

# 3 Linear Neural Networks for Regression

Before we worry about making our neural networks deep, it will be helpful to implement some shallow ones, for which the inputs connect directly to the outputs. This will prove important for a few reasons. First, rather than getting distracted by complicated architectures, we can focus on the basics of neural network training, including parametrizing the output layer, handling data, specifying a loss function, and training the model. Second, this class of shallow networks happens to comprise the set of linear models, which subsumes many classical methods of statistical prediction, including linear and softmax regression. Understanding these classical tools is pivotal because they are widely used in many contexts and we will often need to use them as baselines when justifying the use of fancier architectures. This chapter will focus narrowly on linear regression and the next one will extend our modeling repertoire by developing linear neural networks for classification.

## 3.1 Linear Regression

*Regression* problems pop up whenever we want to predict a numerical value. Common examples include predicting prices (of homes, stocks, etc.), predicting the length of stay (for patients in the hospital), forecasting demand (for retail sales), among numerous others. Not every prediction problem is one of classical regression. Later on, we will introduce classification problems, where the goal is to predict membership among a set of categories.

As a running example, suppose that we wish to estimate the prices of houses (in dollars) based on their area (in square feet) and age (in years). To develop a model for predicting house prices, we need to get our hands on data, including the sales price, area, and age for each home. In the terminology of machine learning, the dataset is called a *training dataset* or *training set*, and each row (containing the data corresponding to one sale) is called an *example* (or *data point*, *instance*, *sample*). The thing we are trying to predict (price) is called a *label* (or *target*). The variables (age and area) upon which the predictions are based are called *features* (or *covariates*).

```
%matplotlib inline
import math
```

```
import time
import numpy as np
import torch
from d2l import torch as d2l
```

## 3.1.1 Basics

*Linear regression* is both the simplest and most popular among the standard tools for tackling regression problems. Dating back to the dawn of the 19th century (Gauss, 1809, Legendre, 1805), linear regression flows from a few simple assumptions. First, we assume that the relationship between features $\mathbf{x}$ and target $y$ is approximately linear, i.e., that the conditional mean $E[Y \mid X = \mathbf{x}]$ can be expressed as a weighted sum of the features $\mathbf{x}$. This setup allows that the target value may still deviate from its expected value on account of observation noise. Next, we can impose the assumption that any such noise is well behaved, following a Gaussian distribution. Typically, we will use $n$ to denote the number of examples in our dataset. We use superscripts to enumerate samples and targets, and subscripts to index coordinates. More concretely, $\mathbf{x}^{(i)}$ denotes the $i^{\textrm{th}}$ sample and $x_j^{(i)}$ denotes its $j^{\textrm{th}}$ coordinate.

### Model

At the heart of every solution is a model that describes how features can be transformed into an estimate of the target. The assumption of linearity means that the expected value of the target (price) can be expressed as a weighted sum of the features (area and age):

$$\textrm{price} = w_{\textrm{area}} \cdot \textrm{area} + w_{\textrm{age}} \cdot \textrm{age} + b. \tag{3.1.1}$$

Here $w_{\textrm{area}}$ and $w_{\textrm{age}}$ are called *weights*, and $b$ is called a *bias* (or *offset* or *intercept*). The weights determine the influence of each feature on our prediction. The bias determines the value of the estimate when all features are zero. Even though we will never see any newly-built homes with precisely zero area, we still need the bias because it allows us to express all linear functions of our features (rather than restricting us to lines that pass through the origin). Strictly speaking, (3.1.1) is an *affine transformation* of input features, which is characterized by a *linear transformation* of features via a weighted sum, combined with a *translation* via the added bias. Given a dataset, our goal is to choose the weights $\mathbf{w}$ and the bias $b$ that, on average, make our model's predictions fit the true prices observed in the data as closely as possible.

In disciplines where it is common to focus on datasets with just a few features, explicitly expressing models long-form, as in (3.1.1), is common. In machine learning, we usually work with high-dimensional datasets, where it is more convenient to employ compact linear algebra notation. When our inputs consist of $d$ features, we can assign each an index (between 1 and $d$) and express our prediction $\hat{y}$ (in general the "hat" symbol denotes an estimate) as

$$\hat{y} = w_1 x_1 + \cdots + w_d x_d + b. \tag{3.1.2}$$

Collecting all features into a vector $\mathbf{x} \in \mathbb{R}^d$ and all weights into a vector $\mathbf{w} \in \mathbb{R}^d$, we can express our model compactly via the dot product between $\mathbf{w}$ and $\mathbf{x}$:

$$\hat{y} = \mathbf{w}^\top \mathbf{x} + b. \tag{3.1.3}$$

In (3.1.3), the vector $\mathbf{x}$ corresponds to the features of a single example. We will often find it convenient to refer to features of our entire dataset of $n$ examples via the *design matrix* $\mathbf{X} \in \mathbb{R}^{n \times d}$. Here, $\mathbf{X}$ contains one row for every example and one column for every feature. For a collection of features $\mathbf{X}$, the predictions $\hat{\mathbf{y}} \in \mathbb{R}^n$ can be expressed via the matrix–vector product:

$$\hat{\mathbf{y}} = \mathbf{X}\mathbf{w} + b, \tag{3.1.4}$$

where broadcasting (Section 2.1.4) is applied during the summation. Given features of a training dataset $\mathbf{X}$ and corresponding (known) labels $\mathbf{y}$, the goal of linear regression is to find the weight vector $\mathbf{w}$ and the bias term $b$ such that, given features of a new data example sampled from the same distribution as $\mathbf{X}$, the new example's label will (in expectation) be predicted with the smallest error.

Even if we believe that the best model for predicting $y$ given $\mathbf{x}$ is linear, we would not expect to find a real-world dataset of $n$ examples where $y^{(i)}$ exactly equals $\mathbf{w}^\top \mathbf{x}^{(i)} + b$ for all $1 \leq i \leq n$. For example, whatever instruments we use to observe the features $\mathbf{X}$ and labels $\mathbf{y}$, there might be a small amount of measurement error. Thus, even when we are confident that the underlying relationship is linear, we will incorporate a noise term to account for such errors.

Before we can go about searching for the best *parameters* (or *model parameters*) $\mathbf{w}$ and $b$, we will need two more things: (i) a measure of the quality of some given model; and (ii) a procedure for updating the model to improve its quality.

## Loss Function

Naturally, fitting our model to the data requires that we agree on some measure of *fitness* (or, equivalently, of *unfitness*). *Loss functions* quantify the distance between the *real* and *predicted* values of the target. The loss will usually be a nonnegative number where smaller values are better and perfect predictions incur a loss of 0. For regression problems, the most common loss function is the squared error. When our prediction for an example $i$ is $\hat{y}^{(i)}$ and the corresponding true label is $y^{(i)}$, the *squared error* is given by:

$$l^{(i)}(\mathbf{w}, b) = \frac{1}{2} \left(\hat{y}^{(i)} - y^{(i)}\right)^2. \tag{3.1.5}$$

The constant $\frac{1}{2}$ makes no real difference but proves to be notationally convenient, since it cancels out when we take the derivative of the loss. Because the training dataset is given to us, and thus is out of our control, the empirical error is only a function of the model parameters. In Fig. 3.1.1, we visualize the fit of a linear regression model in a problem with one-dimensional inputs.

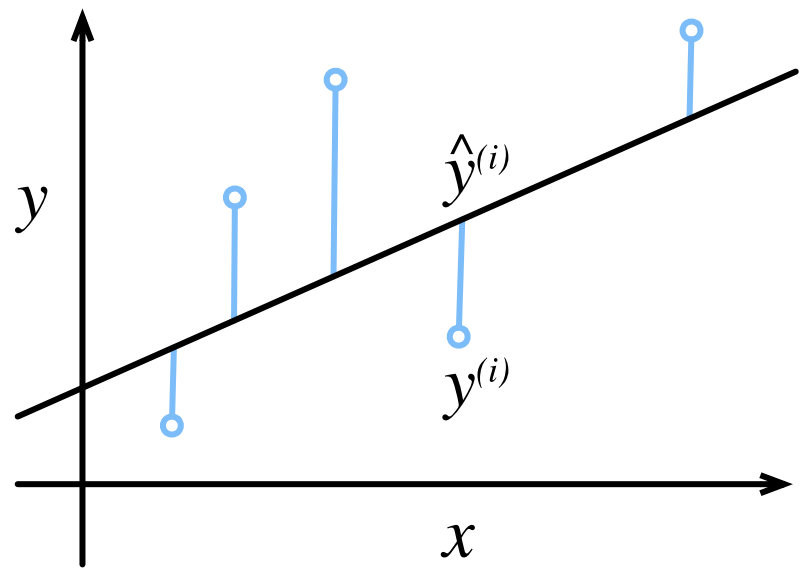

Fig. 3.1.1 Fitting a linear regression model to one-dimensional data.

Note that large differences between estimates $\hat{y}^{(i)}$ and targets $y^{(i)}$ lead to even larger contributions to the loss, due to its quadratic form (this quadraticity can be a double-edge sword; while it encourages the model to avoid large errors it can also lead to excessive sensitivity to anomalous data). To measure the quality of a model on the entire dataset of $n$ examples, we simply average (or equivalently, sum) the losses on the training set:

$$L(\mathbf{w}, b) = \frac{1}{n} \sum_{i=1}^{n} l^{(i)}(\mathbf{w}, b) = \frac{1}{n} \sum_{i=1}^{n} \frac{1}{2} \left(\mathbf{w}^\top \mathbf{x}^{(i)} + b - y^{(i)}\right)^2. \tag{3.1.6}$$

When training the model, we seek parameters $(\mathbf{w}^*, b^*)$ that minimize the total loss across all training examples:

$$\mathbf{w}^*, b^* = \underset{\mathbf{w}, b}{\operatorname{argmin}} \ L(\mathbf{w}, b). \tag{3.1.7}$$

## Analytic Solution

Unlike most of the models that we will cover, linear regression presents us with a surprisingly easy optimization problem. In particular, we can find the optimal parameters (as assessed on the training data) analytically by applying a simple formula as follows. First, we can subsume the bias $b$ into the parameter $\mathbf{w}$ by appending a column to the design matrix consisting of all 1s. Then our prediction problem is to minimize $\|\mathbf{y} - \mathbf{X}\mathbf{w}\|^2$. As long as the design matrix $\mathbf{X}$ has full rank (no feature is linearly dependent on the others), then there will be just one critical point on the loss surface and it corresponds to the minimum of the loss over the entire domain. Taking the derivative of the loss with respect to $\mathbf{w}$ and setting it equal to zero yields:

$$\partial_{\mathbf{w}} \|\mathbf{y} - \mathbf{X}\mathbf{w}\|^2 = 2\mathbf{X}^\top (\mathbf{X}\mathbf{w} - \mathbf{y}) = 0 \text{ and hence } \mathbf{X}^\top \mathbf{y} = \mathbf{X}^\top \mathbf{X}\mathbf{w}. \tag{3.1.8}$$

Solving for $\mathbf{w}$ provides us with the optimal solution for the optimization problem. Note that this solution

$$\mathbf{w}^* = (\mathbf{X}^\top \mathbf{X})^{-1} \mathbf{X}^\top \mathbf{y} \tag{3.1.9}$$

will only be unique when the matrix $\mathbf{X}^\top \mathbf{X}$ is invertible, i.e., when the columns of the design matrix are linearly independent (Golub and Van Loan, 1996).

While simple problems like linear regression may admit analytic solutions, you should not get used to such good fortune. Although analytic solutions allow for nice mathematical analysis, the requirement of an analytic solution is so restrictive that it would exclude almost all exciting aspects of deep learning.

## Minibatch Stochastic Gradient Descent

Fortunately, even in cases where we cannot solve the models analytically, we can still often train models effectively in practice. Moreover, for many tasks, those hard-to-optimize models turn out to be so much better that figuring out how to train them ends up being well worth the trouble.

The key technique for optimizing nearly every deep learning model, and which we will call upon throughout this book, consists of iteratively reducing the error by updating the parameters in the direction that incrementally lowers the loss function. This algorithm is called *gradient descent*.

The most naive application of gradient descent consists of taking the derivative of the loss function, which is an average of the losses computed on every single example in the dataset. In practice, this can be extremely slow: we must pass over the entire dataset before making a single update, even if the update steps might be very powerful (Liu and Nocedal, 1989). Even worse, if there is a lot of redundancy in the training data, the benefit of a full update is limited.

The other extreme is to consider only a single example at a time and to take update steps based on one observation at a time. The resulting algorithm, *stochastic gradient descent* (SGD) can be an effective strategy (Bottou, 2010), even for large datasets. Unfortunately, SGD has drawbacks, both computational and statistical. One problem arises from the fact that processors are a lot faster multiplying and adding numbers than they are at moving data from main memory to processor cache. It is up to an order of magnitude more efficient to perform a matrix–vector multiplication than a corresponding number of vector–vector operations. This means that it can take a lot longer to process one sample at a time compared to a full batch. A second problem is that some of the layers, such as batch normalization (to be described in Section 8.5), only work well when we have access to more than one observation at a time.

The solution to both problems is to pick an intermediate strategy: rather than taking a full batch or only a single sample at a time, we take a *minibatch* of observations (Li *et al.*, 2014). The specific choice of the size of the said minibatch depends on many factors, such as the amount of memory, the number of accelerators, the choice of layers, and the total dataset size. Despite all that, a number between 32 and 256, preferably a multiple of a large power of 2, is a good start. This leads us to *minibatch stochastic gradient descent*.

In its most basic form, in each iteration $t$, we first randomly sample a minibatch $\mathcal{B}_t$ consisting of a fixed number $|\mathcal{B}|$ of training examples. We then compute the derivative (gradient) of the average loss on the minibatch with respect to the model parameters. Finally, we multiply the gradient by a predetermined small positive value $\eta$, called the *learning rate*, and subtract the resulting term from the current parameter values. We can express the update as follows:

$$(\mathbf{w}, b) \leftarrow (\mathbf{w}, b) - \frac{\eta}{|\mathcal{B}|} \sum_{i \in \mathcal{B}_t} \partial_{(\mathbf{w}, b)} l^{(i)}(\mathbf{w}, b). \tag{3.1.10}$$

In summary, minibatch SGD proceeds as follows: (i) initialize the values of the model parameters, typically at random; (ii) iteratively sample random minibatches from the data, updating the parameters in the direction of the negative gradient. For quadratic losses and affine transformations, this has a closed-form expansion:

$$\begin{aligned}
\mathbf{w} &\leftarrow \mathbf{w} - \frac{\eta}{|\mathcal{B}|} \sum_{i \in \mathcal{B}_t} \partial_{\mathbf{w}} l^{(i)}(\mathbf{w}, b) && = \mathbf{w} - \frac{\eta}{|\mathcal{B}|} \sum_{i \in \mathcal{B}_t} \mathbf{x}^{(i)} \left(\mathbf{w}^\top \mathbf{x}^{(i)} + b - y^{(i)}\right)\\
b &\leftarrow b - \frac{\eta}{|\mathcal{B}|} \sum_{i \in \mathcal{B}_t} \partial_b l^{(i)}(\mathbf{w}, b) && = b - \frac{\eta}{|\mathcal{B}|} \sum_{i \in \mathcal{B}_t} \left(\mathbf{w}^\top \mathbf{x}^{(i)} + b - y^{(i)}\right).
\end{aligned} \tag{3.1.11}$$

Since we pick a minibatch $\mathcal{B}$ we need to normalize by its size $|\mathcal{B}|$. Frequently minibatch size and learning rate are user-defined. Such tunable parameters that are not updated in the training loop are called *hyperparameters*. They can be tuned automatically by a number of techniques, such as Bayesian optimization (Frazier, 2018). In the end, the quality of the solution is typically assessed on a separate *validation dataset* (or *validation set*).

After training for some predetermined number of iterations (or until some other stopping criterion is met), we record the estimated model parameters, denoted $\hat{\mathbf{w}}, \hat{b}$. Note that even if our function is truly linear and noiseless, these parameters will not be the exact minimizers of the loss, nor even deterministic. Although the algorithm converges slowly towards the minimizers it typically will not find them exactly in a finite number of steps. Moreover, the minibatches $\mathcal{B}$ used for updating the parameters are chosen at random. This breaks determinism.

Linear regression happens to be a learning problem with a global minimum (whenever $\mathbf{X}$ is full rank, or equivalently, whenever $\mathbf{X}^\top\mathbf{X}$ is invertible). However, the loss surfaces for deep networks contain many saddle points and minima. Fortunately, we typically do not care about finding an exact set of parameters but merely any set of parameters that leads to accurate predictions (and thus low loss). In practice, deep learning practitioners seldom struggle to find parameters that minimize the loss *on training sets* (Frankle and Carbin, 2018, Izmailov *et al.*, 2018). The more formidable task is to find parameters that lead to accurate predictions on previously unseen data, a challenge called *generalization*. We return to these topics throughout the book.

## Predictions

Given the model $\hat{\mathbf{w}}^\top \mathbf{x} + \hat{b}$, we can now make *predictions* for a new example, e.g., predicting the sales price of a previously unseen house given its area $x_1$ and age $x_2$. Deep learning practitioners have taken to calling the prediction phase *inference* but this is a bit of a misnomer—*inference* refers broadly to any conclusion reached on the basis of evidence, including both the values of the parameters and the likely label for an unseen instance. If anything, in the statistics literature *inference* more often denotes parameter inference and this overloading of terminology creates unnecessary confusion when deep learning practitioners talk to statisticians. In the following we will stick to *prediction* whenever possible.

### 3.1.2 Vectorization for Speed

When training our models, we typically want to process whole minibatches of examples simultaneously. Doing this efficiently requires that we vectorize the calculations and leverage fast linear algebra libraries rather than writing costly for-loops in Python.

To see why this matters so much, let's consider two methods for adding vectors. To start, we instantiate two 10,000-dimensional vectors containing all 1s. In the first method, we loop over the vectors with a Python for-loop. In the second, we rely on a single call to +.

```
n = 10000
a = torch.ones(n)
b = torch.ones(n)
```

Now we can benchmark the workloads. First, we add them, one coordinate at a time, using a for-loop.

```
c = torch.zeros(n)
t = time.time()
for i in range(n):
    c[i] = a[i] + b[i]
f'{time.time() - t:.5f} sec'
```

```
'0.16781 sec'
```

Alternatively, we rely on the reloaded + operator to compute the elementwise sum.

```
t = time.time()
d = a + b
f'{time.time() - t:.5f} sec'
```

```
'0.00180 sec'
```

The second method is dramatically faster than the first. Vectorizing code often yields order-of-magnitude speedups. Moreover, we push more of the mathematics to the library so we do not have to write as many calculations ourselves, reducing the potential for errors and increasing portability of the code.

### 3.1.3 The Normal Distribution and Squared Loss

So far we have given a fairly functional motivation of the squared loss objective: the optimal parameters return the conditional expectation $E[Y \mid X]$ whenever the underlying pattern is truly linear, and the loss assigns large penalties for outliers. We can also provide a more formal motivation for the squared loss objective by making probabilistic assumptions about the distribution of noise.

Linear regression was invented at the turn of the 19th century. While it has long been debated whether Gauss or Legendre first thought up the idea, it was Gauss who also discovered the normal distribution (also called the *Gaussian*). It turns out that the normal distribution and linear regression with squared loss share a deeper connection than common parentage.

To begin, recall that a normal distribution with mean $\mu$ and variance $\sigma^2$ (standard deviation $\sigma$) is given as

$$p(x) = \frac{1}{\sqrt{2\pi\sigma^2}} \exp\left(-\frac{1}{2\sigma^2}(x-\mu)^2\right). \tag{3.1.12}$$

Below we define a function to compute the normal distribution.

```
def normal(x, mu, sigma):
    p = 1 / math.sqrt(2 * math.pi * sigma**2)
    return p * np.exp(-0.5 * (x - mu)**2 / sigma**2)
```

We can now visualize the normal distributions.

```
# Use NumPy again for visualization
x = np.arange(-7, 7, 0.01)

# Mean and standard deviation pairs
params = [(0, 1), (0, 2), (3, 1)]
d2l.plot(x, [normal(x, mu, sigma) for mu, sigma in params], xlabel='x',
         ylabel='p(x)', figsize=(4.5, 2.5),
         legend=[f'mean {mu}, std {sigma}' for mu, sigma in params])
```

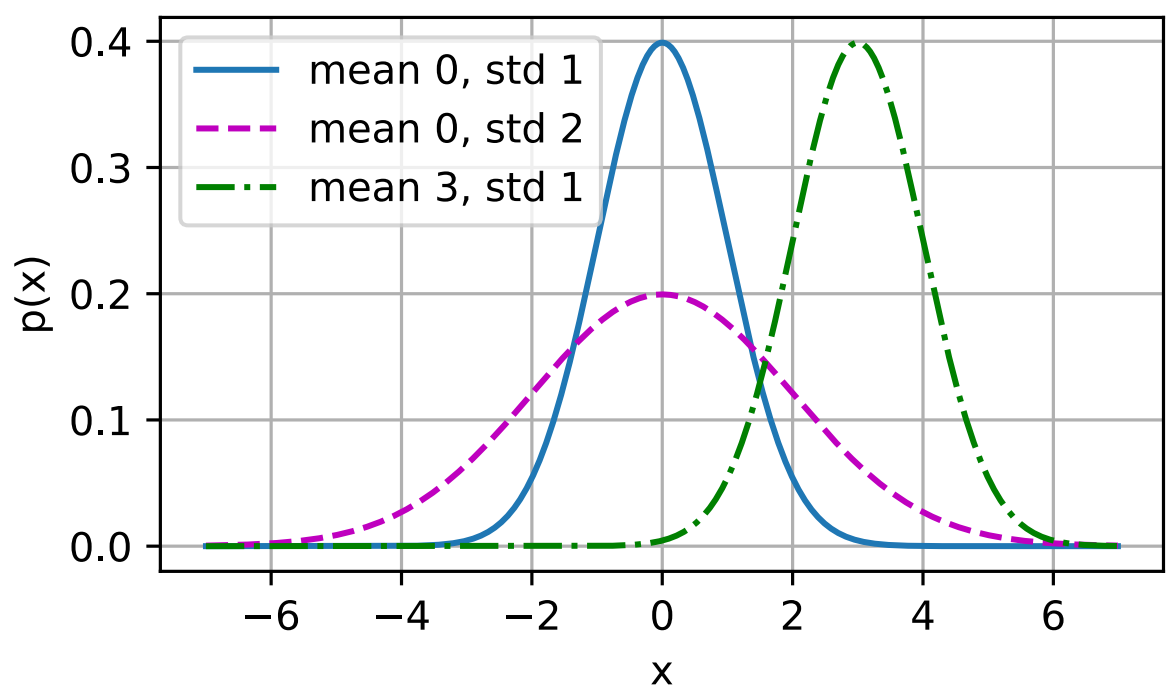

Note that changing the mean corresponds to a shift along the $x$-axis, and increasing the variance spreads the distribution out, lowering its peak.

One way to motivate linear regression with squared loss is to assume that observations arise from noisy measurements, where the noise $\epsilon$ follows the normal distribution $\mathcal{N}(0, \sigma^2)$:

$$y = \mathbf{w}^\top \mathbf{x} + b + \epsilon \text{ where } \epsilon \sim \mathcal{N}(0, \sigma^2). \tag{3.1.13}$$

Thus, we can now write out the *likelihood* of seeing a particular $y$ for a given $\mathbf{x}$ via

$$P(y \mid \mathbf{x}) = \frac{1}{\sqrt{2\pi\sigma^2}} \exp\left(-\frac{1}{2\sigma^2}(y - \mathbf{w}^\top \mathbf{x} - b)^2\right). \tag{3.1.14}$$

As such, the likelihood factorizes. According to *the principle of maximum likelihood*, the best values of parameters $\mathbf{w}$ and $b$ are those that maximize the *likelihood* of the entire dataset:

$$P(\mathbf{y} \mid \mathbf{X}) = \prod_{i=1}^{n} p(y^{(i)} \mid \mathbf{x}^{(i)}). \tag{3.1.15}$$

The equality follows since all pairs $(\mathbf{x}^{(i)}, y^{(i)})$ were drawn independently of each other. Estimators chosen according to the principle of maximum likelihood are called *maximum likelihood estimators*. While, maximizing the product of many exponential functions, might look difficult, we can simplify things significantly, without changing the objective, by maximizing the logarithm of the likelihood instead. For historical reasons, optimizations are more often expressed as minimization rather than maximization. So, without changing anything, we can *minimize* the *negative log-likelihood*, which we can express as follows:

$$-\log P(\mathbf{y} \mid \mathbf{X}) = \sum_{i=1}^{n} \frac{1}{2} \log(2\pi\sigma^2) + \frac{1}{2\sigma^2}\left(y^{(i)} - \mathbf{w}^\top \mathbf{x}^{(i)} - b\right)^2. \tag{3.1.16}$$

If we assume that $\sigma$ is fixed, we can ignore the first term, because it does not depend on $\mathbf{w}$ or $b$. The second term is identical to the squared error loss introduced earlier, except for the multiplicative constant $\frac{1}{\sigma^2}$. Fortunately, the solution does not depend on $\sigma$ either. It follows that minimizing the mean squared error is equivalent to the maximum likelihood estimation of a linear model under the assumption of additive Gaussian noise.

### 3.1.4 Linear Regression as a Neural Network

While linear models are not sufficiently rich to express the many complicated networks that we will introduce in this book, (artificial) neural networks are rich enough to subsume linear models as networks in which every feature is represented by an input neuron, all of which are connected directly to the output.

Fig. 3.1.2 depicts linear regression as a neural network. The diagram highlights the connectivity pattern, such as how each input is connected to the output, but not the specific values taken by the weights or biases.

The inputs are $x_1, \ldots, x_d$. We refer to $d$ as the *number of inputs* or the *feature dimensionality* in the input layer. The output of the network is $o_1$. Because we are just trying to predict a single numerical value, we have only one output neuron. Note that the input values are all *given*. There is just a single *computed* neuron. In summary, we can think of linear regression as a single-layer fully connected neural network. We will encounter networks with far more layers in later chapters.

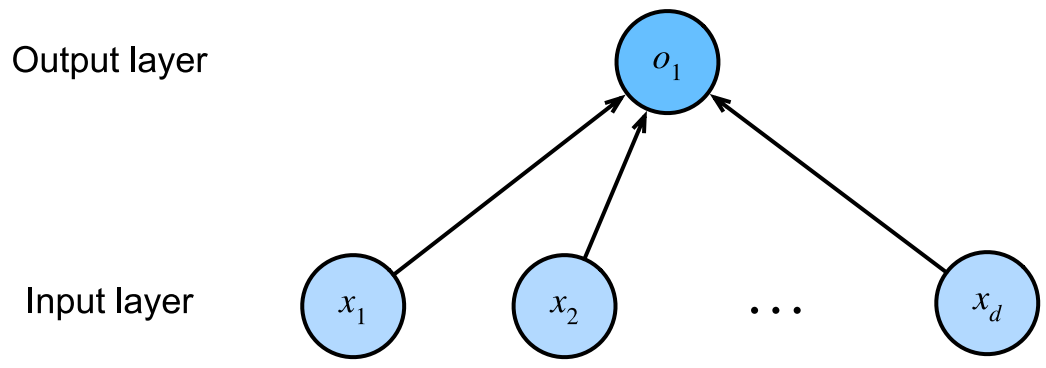

Fig. 3.1.2 Linear regression is a single-layer neural network.

### Biology

Because linear regression predates computational neuroscience, it might seem anachronistic to describe linear regression in terms of neural networks. Nonetheless, they were a natural place to start when the cyberneticists and neurophysiologists Warren McCulloch and Walter Pitts began to develop models of artificial neurons. Consider the cartoonish picture of a biological neuron in Fig. 3.1.3, consisting of *dendrites* (input terminals), the *nucleus* (CPU), the *axon* (output wire), and the *axon terminals* (output terminals), enabling connections to other neurons via *synapses*.



Fig. 3.1.3 The real neuron (source: "Anatomy and Physiology" by the US National Cancer Institute's Surveillance, Epidemiology and End Results (SEER) Program).

Information $x_i$ arriving from other neurons (or environmental sensors) is received in the dendrites. In particular, that information is weighted by *synaptic weights* $w_i$, determining the effect of the inputs, e.g., activation or inhibition via the product $x_i w_i$. The weighted inputs arriving from multiple sources are aggregated in the nucleus as a weighted sum $y = \sum_i x_i w_i + b$, possibly subject to some nonlinear postprocessing via a function $\sigma(y)$. This information is then sent via the axon to the axon terminals, where it reaches its destination (e.g., an actuator such as a muscle) or it is fed into another neuron via its dendrites.

Certainly, the high-level idea that many such units could be combined, provided they have the correct connectivity and learning algorithm, to produce far more interesting and complex behavior than any one neuron alone could express arises from our study of real biological neural systems. At the same time, most research in deep learning today draws inspiration from a much wider source. We invoke Russell and Norvig (2016) who pointed out that although airplanes might have been *inspired* by birds, ornithology has not been the primary driver of aeronautics innovation for some centuries. Likewise, inspiration in deep learning these days comes in equal or greater measure from mathematics, linguistics, psychology, statistics, computer science, and many other fields.

### 3.1.5 Summary

In this section, we introduced traditional linear regression, where the parameters of a linear function are chosen to minimize squared loss on the training set. We also motivated this choice of objective both via some practical considerations and through an interpretation of linear regression as maximum likelihood estimation under an assumption of linearity and Gaussian noise. After discussing both computational considerations and connections to statistics, we showed how such linear models could be expressed as simple neural networks where the inputs are directly wired to the output(s). While we will soon move past linear models altogether, they are sufficient to introduce most of the components that all of our models require: parametric forms, differentiable objectives, optimization via minibatch stochastic gradient descent, and ultimately, evaluation on previously unseen data.

### 3.1.6 Exercises

1. Assume that we have some data $x_1, \ldots, x_n \in \mathbb{R}$. Our goal is to find a constant $b$ such that $\sum_i (x_i - b)^2$ is minimized.
    1. Find an analytic solution for the optimal value of $b$.
    2. How does this problem and its solution relate to the normal distribution?
    3. What if we change the loss from $\sum_i (x_i - b)^2$ to $\sum_i |x_i - b|$? Can you find the optimal solution for $b$?
2. Prove that the affine functions that can be expressed by $\mathbf{x}^\top \mathbf{w} + b$ are equivalent to linear functions on $(\mathbf{x}, 1)$.
3. Assume that you want to find quadratic functions of $\mathbf{x}$, i.e., $f(\mathbf{x}) = b + \sum_i w_i x_i + \sum_{j \leq i} w_{ij} x_i x_j$. How would you formulate this in a deep network?
4. Recall that one of the conditions for the linear regression problem to be solvable was that the design matrix $\mathbf{X}^\top \mathbf{X}$ has full rank.
    1. What happens if this is not the case?
    2. How could you fix it? What happens if you add a small amount of coordinate-wise independent Gaussian noise to all entries of $\mathbf{X}$?
    3. What is the expected value of the design matrix $\mathbf{X}^\top \mathbf{X}$ in this case?
    4. What happens with stochastic gradient descent when $\mathbf{X}^\top \mathbf{X}$ does not have full rank?
5. Assume that the noise model governing the additive noise $\epsilon$ is the exponential distribution. That is, $p(\epsilon) = \frac{1}{2} \exp(-|\epsilon|)$.

1. Write out the negative log-likelihood of the data under the model $-\log P(\mathbf{y} \mid \mathbf{X})$.
   2. Can you find a closed form solution?
   3. Suggest a minibatch stochastic gradient descent algorithm to solve this problem. What could possibly go wrong (hint: what happens near the stationary point as we keep on updating the parameters)? Can you fix this?

6. Assume that we want to design a neural network with two layers by composing two linear layers. That is, the output of the first layer becomes the input of the second layer. Why would such a naive composition not work?
7. What happens if you want to use regression for realistic price estimation of houses or stock prices?
   1. Show that the additive Gaussian noise assumption is not appropriate. Hint: can we have negative prices? What about fluctuations?
   2. Why would regression to the logarithm of the price be much better, i.e., $y = \log \text{price}$?
   3. What do you need to worry about when dealing with pennystock, i.e., stock with very low prices? Hint: can you trade at all possible prices? Why is this a bigger problem for cheap stock? For more information review the celebrated Black–Scholes model for option pricing (Black and Scholes, 1973).
8. Suppose we want to use regression to estimate the *number* of apples sold in a grocery store.
   1. What are the problems with a Gaussian additive noise model? Hint: you are selling apples, not oil.
   2. The Poisson distribution[68] captures distributions over counts. It is given by $p(k \mid \lambda) = \lambda^k e^{-\lambda}/k!$. Here $\lambda$ is the rate function and $k$ is the number of events you see. Prove that $\lambda$ is the expected value of counts $k$.
   3. Design a loss function associated with the Poisson distribution.
   4. Design a loss function for estimating $\log \lambda$ instead.

68 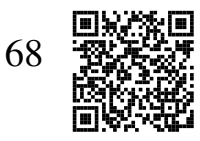

Discussions[69].

69 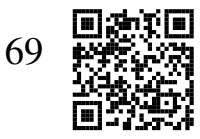

## 3.2 Object-Oriented Design for Implementation

In our introduction to linear regression, we walked through various components including the data, the model, the loss function, and the optimization algorithm. Indeed, linear regression is one of the simplest machine learning models. Training it, however, uses many of the same components that other models in this book require. Therefore, before diving into the implementation details it is worth designing some of the APIs that we use throughout. Treating components in deep learning as objects, we can start by defining classes for these objects and their interactions. This object-oriented design for implementation will greatly streamline the presentation and you might even want to use it in your projects.

Inspired by open-source libraries such as PyTorch Lightning[70], at a high level we wish to have three classes: (i) `Module` contains models, losses, and optimization methods; (ii) `DataModule` provides data loaders for training and validation; (iii) both classes are combined using the `Trainer` class, which allows us to train models on a variety of hardware platforms. Most code in this book adapts `Module` and `DataModule`. We will touch upon the `Trainer` class only when we discuss GPUs, CPUs, parallel training, and optimization algorithms.

70 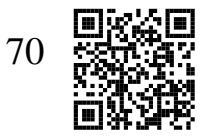

```
import time
import numpy as np
import torch
from torch import nn
from d2l import torch as d2l
```

### 3.2.1 Utilities

We need a few utilities to simplify object-oriented programming in Jupyter notebooks. One of the challenges is that class definitions tend to be fairly long blocks of code. Notebook readability demands short code fragments, interspersed with explanations, a requirement incompatible with the style of programming common for Python libraries. The first utility function allows us to register functions as methods in a class *after* the class has been created. In fact, we can do so *even after* we have created instances of the class! It allows us to split the implementation of a class into multiple code blocks.

```
def add_to_class(Class):  #@save
    """Register functions as methods in created class."""
    def wrapper(obj):
        setattr(Class, obj.__name__, obj)
    return wrapper
```

Let's have a quick look at how to use it. We plan to implement a class `A` with a method `do`. Instead of having code for both `A` and `do` in the same code block, we can first declare the class `A` and create an instance `a`.

```
class A:
    def __init__(self):
        self.b = 1

a = A()
```

Next we define the method `do` as we normally would, but not in class `A`'s scope. Instead, we decorate this method by `add_to_class` with class `A` as its argument. In doing so, the method is able to access the member variables of `A` just as we would expect had it been included as part of `A`'s definition. Let's see what happens when we invoke it for the instance `a`.

```
@add_to_class(A)
def do(self):
    print('Class attribute "b" is', self.b)

a.do()
```

```
Class attribute "b" is 1
```

The second one is a utility class that saves all arguments in a class's `__init__` method as class attributes. This allows us to extend constructor call signatures implicitly without additional code.

```
class HyperParameters:  #@save
    """The base class of hyperparameters."""
    def save_hyperparameters(self, ignore=[]):
        raise NotImplemented
```

We defer its implementation into Section B.7. To use it, we define our class that inherits from `HyperParameters` and calls `save_hyperparameters` in the `__init__` method.

```
# Call the fully implemented HyperParameters class saved in d2l
class B(d2l.HyperParameters):
    def __init__(self, a, b, c):
        self.save_hyperparameters(ignore=['c'])
        print('self.a =', self.a, 'self.b =', self.b)
        print('There is no self.c =', not hasattr(self, 'c'))

b = B(a=1, b=2, c=3)
```

```
self.a = 1 self.b = 2
There is no self.c = True
```

The final utility allows us to plot experiment progress interactively while it is going on. In deference to the much more powerful (and complex) TensorBoard[71] we name it `ProgressBoard`. The implementation is deferred to Section B.7. For now, let's simply see it in action.

71

The `draw` method plots a point `(x, y)` in the figure, with `label` specified in the legend. The optional `every_n` smooths the line by only showing $1/n$ points in the figure. Their values are averaged from the $n$ neighbor points in the original figure.

```
class ProgressBoard(d2l.HyperParameters):  #@save
    """The board that plots data points in animation."""
    def __init__(self, xlabel=None, ylabel=None, xlim=None,
                 ylim=None, xscale='linear', yscale='linear',
```

```
                 ls=['-', '--', '-.', ':'], colors=['C0', 'C1', 'C2', 'C3'],
                 fig=None, axes=None, figsize=(3.5, 2.5), display=True):
        self.save_hyperparameters()

    def draw(self, x, y, label, every_n=1):
        raise NotImplemented
```

In the following example, we draw `sin` and `cos` with a different smoothness. If you run this code block, you will see the lines grow in animation.

```
board = d2l.ProgressBoard('x')
for x in np.arange(0, 10, 0.1):
    board.draw(x, np.sin(x), 'sin', every_n=2)
    board.draw(x, np.cos(x), 'cos', every_n=10)
```

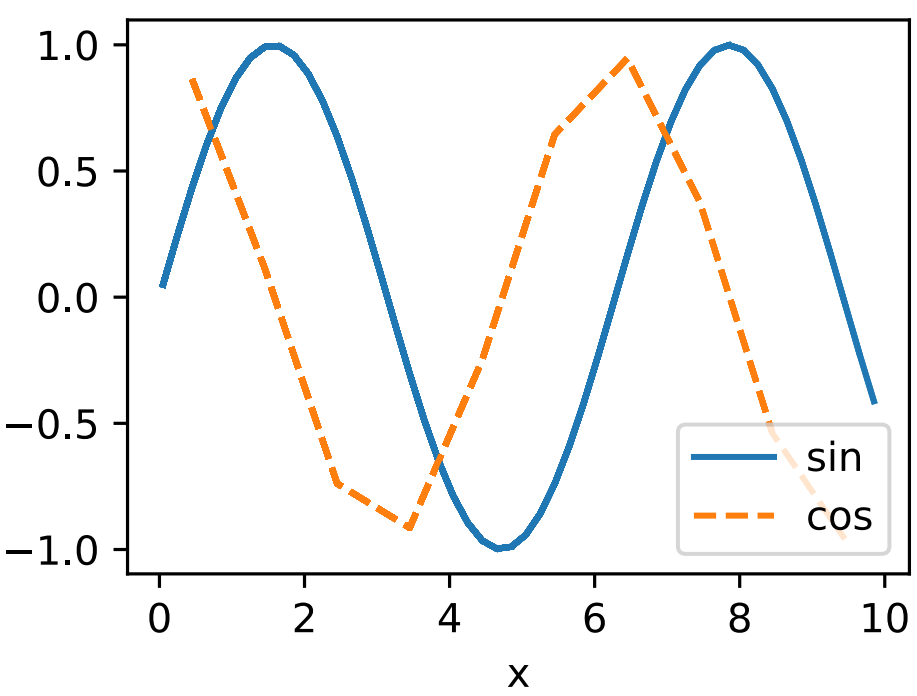

## 3.2.2 Models

The `Module` class is the base class of all models we will implement. At the very least we need three methods. The first, `__init__`, stores the learnable parameters, the `training_step` method accepts a data batch to return the loss value, and finally, `configure_optimizers` returns the optimization method, or a list of them, that is used to update the learnable parameters. Optionally we can define `validation_step` to report the evaluation measures. Sometimes we put the code for computing the output into a separate `forward` method to make it more reusable.

```
class Module(nn.Module, d2l.HyperParameters):  #@save
    """The base class of models."""
    def __init__(self, plot_train_per_epoch=2, plot_valid_per_epoch=1):
        super().__init__()
        self.save_hyperparameters()
        self.board = ProgressBoard()

    def loss(self, y_hat, y):
```

```
        raise NotImplementedError

    def forward(self, X):
        assert hasattr(self, 'net'), 'Neural network is defined'
        return self.net(X)

    def plot(self, key, value, train):
        """Plot a point in animation."""
        assert hasattr(self, 'trainer'), 'Trainer is not inited'
        self.board.xlabel = 'epoch'
        if train:
            x = self.trainer.train_batch_idx / \
                self.trainer.num_train_batches
            n = self.trainer.num_train_batches / \
                self.plot_train_per_epoch
        else:
            x = self.trainer.epoch + 1
            n = self.trainer.num_val_batches / \
                self.plot_valid_per_epoch
        self.board.draw(x, value.to(d2l.cpu()).detach().numpy(),
                        ('train_' if train else 'val_') + key,
                        every_n=int(n))

    def training_step(self, batch):
        l = self.loss(self(*batch[:-1]), batch[-1])
        self.plot('loss', l, train=True)
        return l

    def validation_step(self, batch):
        l = self.loss(self(*batch[:-1]), batch[-1])
        self.plot('loss', l, train=False)

    def configure_optimizers(self):
        raise NotImplementedError
```

You may notice that `Module` is a subclass of `nn.Module`, the base class of neural networks in PyTorch. It provides convenient features for handling neural networks. For example, if we define a `forward` method, such as `forward(self, X)`, then for an instance `a` we can invoke this method by `a(X)`. This works since it calls the `forward` method in the built-in `__call__` method. You can find more details and examples about `nn.Module` in Section 6.1.

### 3.2.3 Data

The `DataModule` class is the base class for data. Quite frequently the `__init__` method is used to prepare the data. This includes downloading and preprocessing if needed. The `train_dataloader` returns the data loader for the training dataset. A data loader is a (Python) generator that yields a data batch each time it is used. This batch is then fed into the `training_step` method of `Module` to compute the loss. There is an optional `val_dataloader` to return the validation dataset loader. It behaves in the same manner, except that it yields data batches for the `validation_step` method in `Module`.

```
class DataModule(d2l.HyperParameters):  #@save
    """The base class of data."""
    def __init__(self, root='../data', num_workers=4):
        self.save_hyperparameters()

    def get_dataloader(self, train):
        raise NotImplementedError

    def train_dataloader(self):
        return self.get_dataloader(train=True)

    def val_dataloader(self):
        return self.get_dataloader(train=False)
```

### 3.2.4 Training

The `Trainer` class trains the learnable parameters in the `Module` class with data specified in `DataModule`. The key method is `fit`, which accepts two arguments: `model`, an instance of `Module`, and `data`, an instance of `DataModule`. It then iterates over the entire dataset `max_epochs` times to train the model. As before, we will defer the implementation of this method to later chapters.

```
class Trainer(d2l.HyperParameters):  #@save
    """The base class for training models with data."""
    def __init__(self, max_epochs, num_gpus=0, gradient_clip_val=0):
        self.save_hyperparameters()
        assert num_gpus == 0, 'No GPU support yet'

    def prepare_data(self, data):
        self.train_dataloader = data.train_dataloader()
        self.val_dataloader = data.val_dataloader()
        self.num_train_batches = len(self.train_dataloader)
        self.num_val_batches = (len(self.val_dataloader)
                                if self.val_dataloader is not None else 0)

    def prepare_model(self, model):
        model.trainer = self
        model.board.xlim = [0, self.max_epochs]
        self.model = model

    def fit(self, model, data):
        self.prepare_data(data)
        self.prepare_model(model)
        self.optim = model.configure_optimizers()
        self.epoch = 0
        self.train_batch_idx = 0
        self.val_batch_idx = 0
        for self.epoch in range(self.max_epochs):
            self.fit_epoch()

    def fit_epoch(self):
        raise NotImplementedError
```

### 3.2.5 Summary

To highlight the object-oriented design for our future deep learning implementation, the above classes simply show how their objects store data and interact with each other. We will keep enriching implementations of these classes, such as via `@add_to_class`, in the rest of the book. Moreover, these fully implemented classes are saved in the D2L library[72], a *lightweight toolkit* that makes structured modeling for deep learning easy. In particular, it facilitates reusing many components between projects without changing much at all. For instance, we can replace just the optimizer, just the model, just the dataset, etc.; this degree of modularity pays dividends throughout the book in terms of conciseness and simplicity (this is why we added it) and it can do the same for your own projects.

### 3.2.6 Exercises

1. Locate full implementations of the above classes that are saved in the D2L library[73]. We strongly recommend that you look at the implementation in detail once you have gained some more familiarity with deep learning modeling.
2. Remove the `save_hyperparameters` statement in the `B` class. Can you still print `self.a` and `self.b`? Optional: if you have dived into the full implementation of the `HyperParameters` class, can you explain why?

Discussions[74].

## 3.3 Synthetic Regression Data

Machine learning is all about extracting information from data. So you might wonder, what could we possibly learn from synthetic data? While we might not care intrinsically about the patterns that we ourselves baked into an artificial data generating model, such datasets are nevertheless useful for didactic purposes, helping us to evaluate the properties of our learning algorithms and to confirm that our implementations work as expected. For example, if we create data for which the correct parameters are known *a priori*, then we can check that our model can in fact recover them.

```
%matplotlib inline
import random
import torch
from d2l import torch as d2l
```

### 3.3.1 Generating the Dataset

For this example, we will work in low dimension for succinctness. The following code snippet generates 1000 examples with 2-dimensional features drawn from a standard normal distribution. The resulting design matrix $\mathbf{X}$ belongs to $\mathbb{R}^{1000\times 2}$. We generate each label by applying a *ground truth* linear function, corrupting them via additive noise $\boldsymbol{\epsilon}$, drawn independently and identically for each example:

$$\mathbf{y} = \mathbf{X}\mathbf{w} + b + \boldsymbol{\epsilon}. \tag{3.3.1}$$

For convenience we assume that $\boldsymbol{\epsilon}$ is drawn from a normal distribution with mean $\mu = 0$ and standard deviation $\sigma = 0.01$. Note that for object-oriented design we add the code to the `__init__` method of a subclass of `d2l.DataModule` (introduced in Section 3.2.3). It is good practice to allow the setting of any additional hyperparameters. We accomplish this with `save_hyperparameters()`. The `batch_size` will be determined later.

```
class SyntheticRegressionData(d2l.DataModule):  #@save
    """Synthetic data for linear regression."""
    def __init__(self, w, b, noise=0.01, num_train=1000, num_val=1000,
                 batch_size=32):
        super().__init__()
        self.save_hyperparameters()
        n = num_train + num_val
        self.X = torch.randn(n, len(w))
        noise = torch.randn(n, 1) * noise
        self.y = torch.matmul(self.X, w.reshape((-1, 1))) + b + noise
```

Below, we set the true parameters to $\mathbf{w} = [2, -3.4]^\top$ and $b = 4.2$. Later, we can check our estimated parameters against these *ground truth* values.

```
data = SyntheticRegressionData(w=torch.tensor([2, -3.4]), b=4.2)
```

Each row in `features` consists of a vector in $\mathbb{R}^2$ and each row in `labels` is a scalar. Let's have a look at the first entry.

```
print('features:', data.X[0],'\nlabel:', data.y[0])
```

```
features: tensor([ 2.2793, -0.2246])
label: tensor([9.5014])
```

## 3.3.2 Reading the Dataset

Training machine learning models often requires multiple passes over a dataset, grabbing one minibatch of examples at a time. This data is then used to update the model. To illustrate how this works, we implement the `get_dataloader` method, registering it in the `SyntheticRegressionData` class via `add_to_class` (introduced in Section 3.2.1). It takes a batch size, a matrix of features, and a vector of labels, and generates minibatches of size `batch_size`.

As such, each minibatch consists of a tuple of features and labels. Note that we need to be mindful of whether we're in training or validation mode: in the former, we will want to read the data in random order, whereas for the latter, being able to read data in a pre-defined order may be important for debugging purposes.

```
@d2l.add_to_class(SyntheticRegressionData)
def get_dataloader(self, train):
    if train:
        indices = list(range(0, self.num_train))
        # The examples are read in random order
        random.shuffle(indices)
    else:
        indices = list(range(self.num_train, self.num_train+self.num_val))
    for i in range(0, len(indices), self.batch_size):
        batch_indices = torch.tensor(indices[i: i+self.batch_size])
        yield self.X[batch_indices], self.y[batch_indices]
```

To build some intuition, let's inspect the first minibatch of data. Each minibatch of features provides us with both its size and the dimensionality of input features. Likewise, our minibatch of labels will have a matching shape given by `batch_size`.

```
X, y = next(iter(data.train_dataloader()))
print('X shape:', X.shape, '\ny shape:', y.shape)
```

```
X shape: torch.Size([32, 2])
y shape: torch.Size([32, 1])
```

While seemingly innocuous, the invocation of `iter(data.train_dataloader())` illustrates the power of Python's object-oriented design. Note that we added a method to the `SyntheticRegressionData` class *after* creating the `data` object. Nonetheless, the object benefits from the *ex post facto* addition of functionality to the class.

Throughout the iteration we obtain distinct minibatches until the entire dataset has been exhausted (try this). While the iteration implemented above is good for didactic purposes, it is inefficient in ways that might get us into trouble with real problems. For example, it requires that we load all the data in memory and that we perform lots of random memory access. The built-in iterators implemented in a deep learning framework are considerably more efficient and they can deal with sources such as data stored in files, data received via a stream, and data generated or processed on the fly. Next let's try to implement the same method using built-in iterators.

### 3.3.3 Concise Implementation of the Data Loader

Rather than writing our own iterator, we can call the existing API in a framework to load data. As before, we need a dataset with features `X` and labels `y`. Beyond that, we set `batch_size` in the built-in data loader and let it take care of shuffling examples efficiently.

```
@d2l.add_to_class(d2l.DataModule)  #@save
def get_tensorloader(self, tensors, train, indices=slice(0, None)):
    tensors = tuple(a[indices] for a in tensors)
    dataset = torch.utils.data.TensorDataset(*tensors)
    return torch.utils.data.DataLoader(dataset, self.batch_size,
                                       shuffle=train)
```

```
@d2l.add_to_class(SyntheticRegressionData)  #@save
def get_dataloader(self, train):
    i = slice(0, self.num_train) if train else slice(self.num_train, None)
    return self.get_tensorloader((self.X, self.y), train, i)
```

The new data loader behaves just like the previous one, except that it is more efficient and has some added functionality.

```
X, y = next(iter(data.train_dataloader()))
print('X shape:', X.shape, '\ny shape:', y.shape)
```

```
X shape: torch.Size([32, 2])
y shape: torch.Size([32, 1])
```

For instance, the data loader provided by the framework API supports the built-in `__len__` method, so we can query its length, i.e., the number of batches.

```
len(data.train_dataloader())
```

```
32
```

### 3.3.4 Summary

Data loaders are a convenient way of abstracting out the process of loading and manipulating data. This way the same machine learning *algorithm* is capable of processing many different types and sources of data without the need for modification. One of the nice things about data loaders is that they can be composed. For instance, we might be loading images and then have a postprocessing filter that crops them or modifies them in other ways. As such, data loaders can be used to describe an entire data processing pipeline.

As for the model itself, the two-dimensional linear model is about the simplest we might encounter. It lets us test out the accuracy of regression models without worrying about having insufficient amounts of data or an underdetermined system of equations. We will put this to good use in the next section.

### 3.3.5 Exercises

1. What will happen if the number of examples cannot be divided by the batch size. How would you change this behavior by specifying a different argument by using the framework's API?
2. Suppose that we want to generate a huge dataset, where both the size of the parameter vector `w` and the number of examples `num_examples` are large.
    1. What happens if we cannot hold all data in memory?
    2. How would you shuffle the data if it is held on disk? Your task is to design an *efficient* algorithm that does not require too many random reads or writes. Hint: pseudorandom permutation generators[75] allow you to design a reshuffle without the need to store the permutation table explicitly (Naor and Reingold, 1999).

75 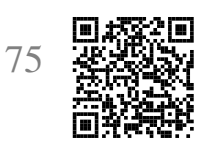

3. Implement a data generator that produces new data on the fly, every time the iterator is called.
4. How would you design a random data generator that generates *the same* data each time it is called?

Discussions[76].

76

## 3.4 Linear Regression Implementation from Scratch

We are now ready to work through a fully functioning implementation of linear regression. In this section, we will implement the entire method from scratch, including (i) the model; (ii) the loss function; (iii) a minibatch stochastic gradient descent optimizer; and (iv) the training function that stitches all of these pieces together. Finally, we will run our synthetic data generator from Section 3.3 and apply our model on the resulting dataset. While modern deep learning frameworks can automate nearly all of this work, implementing things from scratch is the only way to make sure that you really know what you are doing. Moreover, when it is time to customize models, defining our own layers or loss functions, understanding how things work under the hood will prove handy. In this section, we will rely only on tensors and automatic differentiation. Later, we will introduce a more concise implementation, taking advantage of the bells and whistles of deep learning frameworks while retaining the structure of what follows below.

```
%matplotlib inline
import torch
from d2l import torch as d2l
```

### 3.4.1 Defining the Model

Before we can begin optimizing our model's parameters by minibatch SGD, we need to have some parameters in the first place. In the following we initialize weights by drawing random numbers from a normal distribution with mean 0 and a standard deviation of 0.01. The magic number 0.01 often works well in practice, but you can specify a different value through the argument `sigma`. Moreover we set the bias to 0. Note that for object-oriented design we add the code to the `__init__` method of a subclass of `d2l.Module` (introduced in Section 3.2.2).

```
class LinearRegressionScratch(d2l.Module):  #@save
    """The linear regression model implemented from scratch."""
    def __init__(self, num_inputs, lr, sigma=0.01):
        super().__init__()
        self.save_hyperparameters()
        self.w = torch.normal(0, sigma, (num_inputs, 1), requires_grad=True)
        self.b = torch.zeros(1, requires_grad=True)
```

Next we must define our model, relating its input and parameters to its output. Using the same notation as (3.1.4) for our linear model we simply take the matrix–vector product of the input features $\mathbf{X}$ and the model weights $\mathbf{w}$, and add the offset $b$ to each example. The product $\mathbf{Xw}$ is a vector and $b$ is a scalar. Because of the broadcasting mechanism (see Section 2.1.4), when we add a vector and a scalar, the scalar is added to each component of the vector. The resulting `forward` method is registered in the `LinearRegressionScratch` class via `add_to_class` (introduced in Section 3.2.1).

```
@d2l.add_to_class(LinearRegressionScratch)  #@save
def forward(self, X):
    return torch.matmul(X, self.w) + self.b
```

## 3.4.2 Defining the Loss Function

Since updating our model requires taking the gradient of our loss function, we ought to define the loss function first. Here we use the squared loss function in (3.1.5). In the implementation, we need to transform the true value `y` into the predicted value's shape `y_hat`. The result returned by the following method will also have the same shape as `y_hat`. We also return the averaged loss value among all examples in the minibatch.

```
@d2l.add_to_class(LinearRegressionScratch)  #@save
def loss(self, y_hat, y):
    l = (y_hat - y) ** 2 / 2
    return l.mean()
```

## 3.4.3 Defining the Optimization Algorithm

As discussed in Section 3.1, linear regression has a closed-form solution. However, our goal here is to illustrate how to train more general neural networks, and that requires that we teach you how to use minibatch SGD. Hence we will take this opportunity to introduce your first working example of SGD. At each step, using a minibatch randomly drawn from our dataset, we estimate the gradient of the loss with respect to the parameters. Next, we update the parameters in the direction that may reduce the loss.

The following code applies the update, given a set of parameters, a learning rate `lr`. Since our loss is computed as an average over the minibatch, we do not need to adjust the learning rate against the batch size. In later chapters we will investigate how learning rates should be adjusted for very large minibatches as they arise in distributed large-scale learning. For now, we can ignore this dependency.

We define our `SGD` class, a subclass of `d2l.HyperParameters` (introduced in Section 3.2.1), to have a similar API as the built-in SGD optimizer. We update the parameters in the `step` method. The `zero_grad` method sets all gradients to 0, which must be run before a backpropagation step.

```
class SGD(d2l.HyperParameters):  #@save
    """Minibatch stochastic gradient descent."""
    def __init__(self, params, lr):
        self.save_hyperparameters()

    def step(self):
        for param in self.params:
            param -= self.lr * param.grad

    def zero_grad(self):
        for param in self.params:
            if param.grad is not None:
                param.grad.zero_()
```

We next define the `configure_optimizers` method, which returns an instance of the `SGD` class.

```
@d2l.add_to_class(LinearRegressionScratch)  #@save
def configure_optimizers(self):
    return SGD([self.w, self.b], self.lr)
```

## 3.4.4 Training

Now that we have all of the parts in place (parameters, loss function, model, and optimizer), we are ready to implement the main training loop. It is crucial that you understand this code fully since you will employ similar training loops for every other deep learning model covered in this book. In each *epoch*, we iterate through the entire training dataset, passing once through every example (assuming that the number of examples is divisible by the batch size). In each *iteration*, we grab a minibatch of training examples, and compute its loss through the model's `training_step` method. Then we compute the gradients with respect to each parameter. Finally, we will call the optimization algorithm to update the model parameters. In summary, we will execute the following loop:

- Initialize parameters $(\mathbf{w}, b)$
- Repeat until done
  - Compute gradient $\mathbf{g} \leftarrow \partial_{(\mathbf{w},b)} \frac{1}{|\mathcal{B}|} \sum_{i \in \mathcal{B}} l(\mathbf{x}^{(i)}, y^{(i)}, \mathbf{w}, b)$
  - Update parameters $(\mathbf{w}, b) \leftarrow (\mathbf{w}, b) - \eta \mathbf{g}$

Recall that the synthetic regression dataset that we generated in Section 3.3 does not provide a validation dataset. In most cases, however, we will want a validation dataset to measure our model quality. Here we pass the validation dataloader once in each epoch to measure the model performance. Following our object-oriented design, the `prepare_batch` and `fit_epoch` methods are registered in the `d2l.Trainer` class (introduced in Section 3.2.4).

```
@d2l.add_to_class(d2l.Trainer)  #@save
def prepare_batch(self, batch):
    return batch
```

```
@d2l.add_to_class(d2l.Trainer)  #@save
def fit_epoch(self):
    self.model.train()
    for batch in self.train_dataloader:
        loss = self.model.training_step(self.prepare_batch(batch))
        self.optim.zero_grad()
        with torch.no_grad():
            loss.backward()
            if self.gradient_clip_val > 0:  # To be discussed later
                self.clip_gradients(self.gradient_clip_val, self.model)
            self.optim.step()
        self.train_batch_idx += 1
    if self.val_dataloader is None:
        return
    self.model.eval()
    for batch in self.val_dataloader:
        with torch.no_grad():
            self.model.validation_step(self.prepare_batch(batch))
        self.val_batch_idx += 1
```

We are almost ready to train the model, but first we need some training data. Here we use the `SyntheticRegressionData` class and pass in some ground truth parameters. Then we train our model with the learning rate `lr=0.03` and set `max_epochs=3`. Note that in general, both the number of epochs and the learning rate are hyperparameters. In general, setting hyperparameters is tricky and we will usually want to use a three-way split, one set for training, a second for hyperparameter selection, and the third reserved for the final evaluation. We elide these details for now but will revise them later.

```
model = LinearRegressionScratch(2, lr=0.03)
data = d2l.SyntheticRegressionData(w=torch.tensor([2, -3.4]), b=4.2)
trainer = d2l.Trainer(max_epochs=3)
trainer.fit(model, data)
```

Because we synthesized the dataset ourselves, we know precisely what the true parameters are. Thus, we can evaluate our success in training by comparing the true parameters with those that we learned through our training loop. Indeed they turn out to be very close to each other.

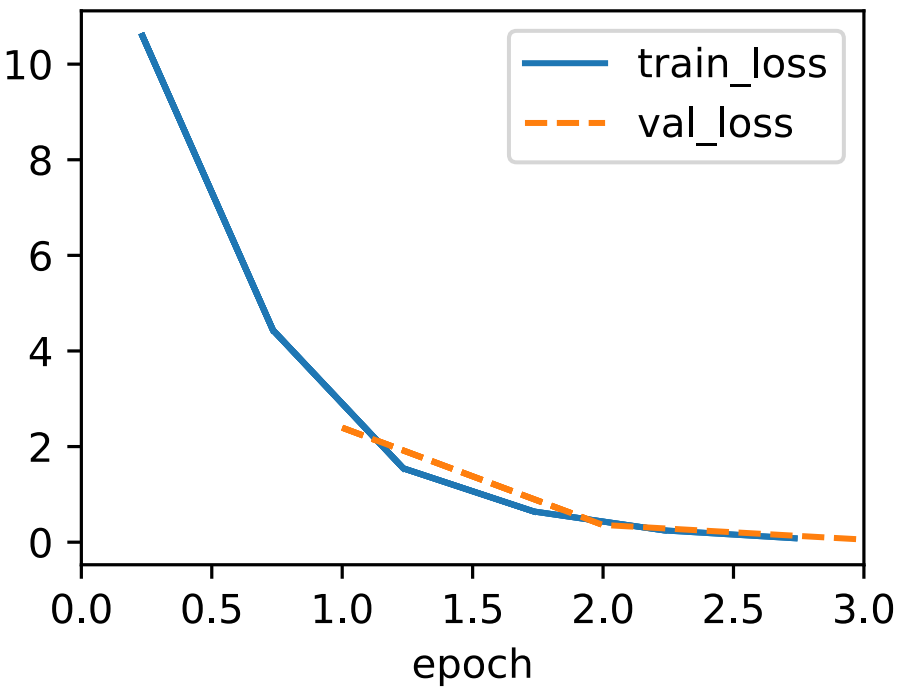

```
with torch.no_grad():
    print(f'error in estimating w: {data.w - model.w.reshape(data.w.shape)}')
    print(f'error in estimating b: {data.b - model.b}')
```

```
error in estimating w: tensor([ 0.1163, -0.2020])
error in estimating b: tensor([0.2383])
```

We should not take the ability to exactly recover the ground truth parameters for granted. In general, for deep models unique solutions for the parameters do not exist, and even for linear models, exactly recovering the parameters is only possible when no feature is linearly dependent on the others. However, in machine learning, we are often less concerned with recovering true underlying parameters, but rather with parameters that lead to highly accurate prediction (Vapnik, 1992). Fortunately, even on difficult optimization problems, stochastic gradient descent can often find remarkably good solutions, owing partly to the fact that, for deep networks, there exist many configurations of the parameters that lead to highly accurate prediction.

### 3.4.5 Summary

In this section, we took a significant step towards designing deep learning systems by implementing a fully functional neural network model and training loop. In this process, we built a data loader, a model, a loss function, an optimization procedure, and a visualization and monitoring tool. We did this by composing a Python object that contains all relevant components for training a model. While this is not yet a professional-grade implementation it is perfectly functional and code like this could already help you to solve small problems quickly. In the coming sections, we will see how to do this both *more concisely* (avoiding boilerplate code) and *more efficiently* (using our GPUs to their full potential).

### 3.4.6 Exercises

1. What would happen if we were to initialize the weights to zero. Would the algorithm still work? What if we initialized the parameters with variance $1000$ rather than $0.01$?

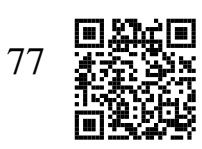

2. Assume that you are Georg Simon Ohm[77] trying to come up with a model for resistance that relates voltage and current. Can you use automatic differentiation to learn the parameters of your model?

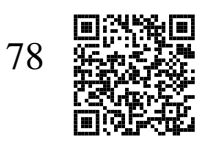

3. Can you use Planck's Law[78] to determine the temperature of an object using spectral energy density? For reference, the spectral density $B$ of radiation emanating from a black body is $B(\lambda, T) = \frac{2hc^2}{\lambda^5} \cdot \left(\exp \frac{hc}{\lambda kT} - 1\right)^{-1}$. Here $\lambda$ is the wavelength, $T$ is the temperature, $c$ is the speed of light, $h$ is Planck's constant, and $k$ is the Boltzmann constant. You measure the energy for different wavelengths $\lambda$ and you now need to fit the spectral density curve to Planck's law.

4. What are the problems you might encounter if you wanted to compute the second derivatives of the loss? How would you fix them?

5. Why is the `reshape` method needed in the `loss` function?

6. Experiment using different learning rates to find out how quickly the loss function value drops. Can you reduce the error by increasing the number of epochs of training?

7. If the number of examples cannot be divided by the batch size, what happens to `data_iter` at the end of an epoch?

8. Try implementing a different loss function, such as the absolute value loss `(y_hat - d2l.reshape(y, y_hat.shape)).abs().sum()`.

    1. Check what happens for regular data.

    2. Check whether there is a difference in behavior if you actively perturb some entries, such as $y_5 = 10000$, of $\mathbf{y}$.

    3. Can you think of a cheap solution for combining the best aspects of squared loss and absolute value loss? Hint: how can you avoid really large gradient values?

9. Why do we need to reshuffle the dataset? Can you design a case where a maliciously constructed dataset would break the optimization algorithm otherwise?

Discussions[79].

## 3.5 Concise Implementation of Linear Regression

Deep learning has witnessed a sort of Cambrian explosion over the past decade. The sheer number of techniques, applications and algorithms by far surpasses the progress of previous decades. This is due to a fortuitous combination of multiple factors, one of which is the powerful free tools offered by a number of open-source deep learning frameworks. Theano (Bergstra *et al.*, 2010), DistBelief (Dean *et al.*, 2012), and Caffe (Jia *et al.*, 2014) arguably represent the first generation of such models that found widespread adoption. In contrast to earlier (seminal) works like SN2 (Simulateur Neuristique) (Bottou and Le Cun, 1988), which provided a Lisp-like programming experience, modern frameworks offer automatic differentiation and the convenience of Python. These frameworks allow us to automate and modularize the repetitive work of implementing gradient-based learning algorithms.

In Section 3.4, we relied only on (i) tensors for data storage and linear algebra; and (ii) automatic differentiation for calculating gradients. In practice, because data iterators, loss functions, optimizers, and neural network layers are so common, modern libraries implement these components for us as well. In this section, we will show you how to implement the linear regression model from Section 3.4 concisely by using high-level APIs of deep learning frameworks.

```
import numpy as np
import torch
from torch import nn
from d2l import torch as d2l
```

### 3.5.1 Defining the Model

When we implemented linear regression from scratch in Section 3.4, we defined our model parameters explicitly and coded up the calculations to produce output using basic linear algebra operations. You *should* know how to do this. But once your models get more complex, and once you have to do this nearly every day, you will be glad of the assistance. The situation is similar to coding up your own blog from scratch. Doing it once or twice is rewarding and instructive, but you would be a lousy web developer if you spent a month reinventing the wheel.

For standard operations, we can use a framework's predefined layers, which allow us to focus on the layers used to construct the model rather than worrying about their implementation. Recall the architecture of a single-layer network as described in Fig. 3.1.2. The layer is called *fully connected*, since each of its inputs is connected to each of its outputs by means of a matrix–vector multiplication.

In PyTorch, the fully connected layer is defined in `Linear` and `LazyLinear` classes (available since version 1.8.0). The latter allows users to specify *merely* the output dimension, while the former additionally asks for how many inputs go into this layer. Specifying input shapes is inconvenient and may require nontrivial calculations (such as in convolutional layers). Thus, for simplicity, we will use such "lazy" layers whenever we can.

```
class LinearRegression(d2l.Module):  #@save
    """The linear regression model implemented with high-level APIs."""
    def __init__(self, lr):
        super().__init__()
        self.save_hyperparameters()
        self.net = nn.LazyLinear(1)
        self.net.weight.data.normal_(0, 0.01)
        self.net.bias.data.fill_(0)
```

In the `forward` method we just invoke the built-in `__call__` method of the predefined layers to compute the outputs.

```
@d2l.add_to_class(LinearRegression)  #@save
def forward(self, X):
    return self.net(X)
```

### 3.5.2 Defining the Loss Function

The `MSELoss` class computes the mean squared error (without the $1/2$ factor in (3.1.5)). By default, `MSELoss` returns the average loss over examples. It is faster (and easier to use) than implementing our own.

```
@d2l.add_to_class(LinearRegression)  #@save
def loss(self, y_hat, y):
    fn = nn.MSELoss()
    return fn(y_hat, y)
```

### 3.5.3 Defining the Optimization Algorithm

Minibatch SGD is a standard tool for optimizing neural networks and thus PyTorch supports it alongside a number of variations on this algorithm in the `optim` module. When we instantiate an `SGD` instance, we specify the parameters to optimize over, obtainable from our model via `self.parameters()`, and the learning rate (`self.lr`) required by our optimization algorithm.

```
@d2l.add_to_class(LinearRegression)  #@save
def configure_optimizers(self):
    return torch.optim.SGD(self.parameters(), self.lr)
```

### 3.5.4 Training

You might have noticed that expressing our model through high-level APIs of a deep learning framework requires fewer lines of code. We did not have to allocate parameters individually, define our loss function, or implement minibatch SGD. Once we start working with much more complex models, the advantages of the high-level API will grow considerably.

Now that we have all the basic pieces in place, the training loop itself is the same as the one we implemented from scratch. So we just call the `fit` method (introduced in Section 3.2.4), which relies on the implementation of the `fit_epoch` method in Section 3.4, to train our model.

```
model = LinearRegression(lr=0.03)
data = d2l.SyntheticRegressionData(w=torch.tensor([2, -3.4]), b=4.2)
trainer = d2l.Trainer(max_epochs=3)
trainer.fit(model, data)
```

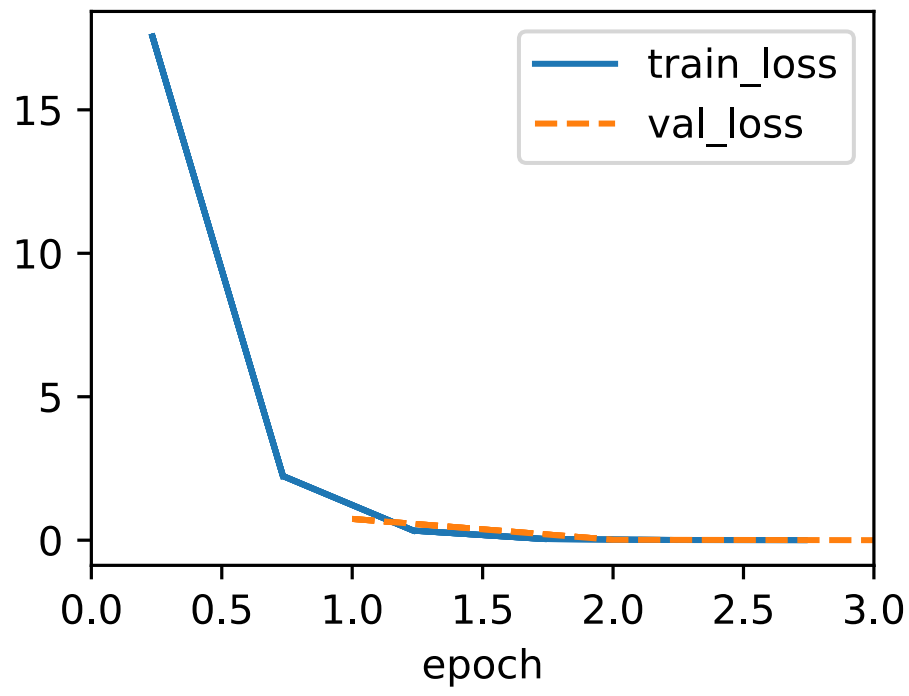

Below, we compare the model parameters learned by training on finite data and the actual parameters that generated our dataset. To access parameters, we access the weights and bias of the layer that we need. As in our implementation from scratch, note that our estimated parameters are close to their true counterparts.

```
@d2l.add_to_class(LinearRegression)  #@save
def get_w_b(self):
    return (self.net.weight.data, self.net.bias.data)
w, b = model.get_w_b()
```

```
print(f'error in estimating w: {data.w - w.reshape(data.w.shape)}')
print(f'error in estimating b: {data.b - b}')
```

```
error in estimating w: tensor([ 0.0031, -0.0099])
error in estimating b: tensor([0.0127])
```

### 3.5.5 Summary

This section contains the first implementation of a deep network (in this book) to tap into the conveniences afforded by modern deep learning frameworks, such as MXNet (Chen *et al.*, 2015), JAX (Frostig *et al.*, 2018), PyTorch (Paszke *et al.*, 2019), and Tensorflow (Abadi *et al.*, 2016). We used framework defaults for loading data, defining a layer, a loss function, an optimizer and a training loop. Whenever the framework provides all necessary features, it is generally a good idea to use them, since the library implementations of these components tend to be heavily optimized for performance and properly tested for reliability. At the same time, try not to forget that these modules *can* be implemented directly. This is especially important for aspiring researchers who wish to live on the leading edge of model development, where you will be inventing new components that cannot possibly exist in any current library.

In PyTorch, the `data` module provides tools for data processing, the `nn` module defines a large number of neural network layers and common loss functions. We can initialize the parameters by replacing their values with methods ending with `_`. Note that we need to specify the input dimensions of the network. While this is trivial for now, it can have significant knock-on effects when we want to design complex networks with many layers. Careful considerations of how to parametrize these networks is needed to allow portability.

### 3.5.6 Exercises

1. How would you need to change the learning rate if you replace the aggregate loss over the minibatch with an average over the loss on the minibatch?
2. Review the framework documentation to see which loss functions are provided. In particular, replace the squared loss with Huber's robust loss function. That is, use the loss function

$$l(y, y') = \begin{cases} |y - y'| - \frac{\sigma}{2} & \text{if } |y - y'| > \sigma \\ \frac{1}{2\sigma}(y - y')^2 & \text{otherwise} \end{cases} \tag{3.5.1}$$

3. How do you access the gradient of the weights of the model?
4. What is the effect on the solution if you change the learning rate and the number of epochs? Does it keep on improving?
5. How does the solution change as you vary the amount of data generated?
    1. Plot the estimation error for $\hat{\mathbf{w}} - \mathbf{w}$ and $\hat{b} - b$ as a function of the amount of data. Hint: increase the amount of data logarithmically rather than linearly, i.e., 5, 10, 20, 50, …, 10,000 rather than 1000, 2000, …, 10,000.

80

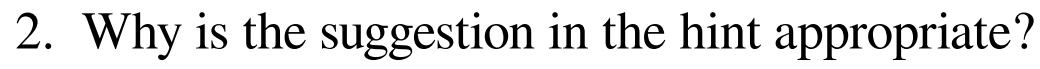

    2. Why is the suggestion in the hint appropriate?

Discussions[80].

## 3.6 Generalization

Consider two college students diligently preparing for their final exam. Commonly, this preparation will consist of practicing and testing their abilities by taking exams administered in previous years. Nonetheless, doing well on past exams is no guarantee that they will excel when it matters. For instance, imagine a student, Extraordinary Ellie, whose preparation consisted entirely of memorizing the answers to previous years' exam questions. Even if Ellie were endowed with an extraordinary memory, and thus could perfectly recall the answer to any *previously seen* question, she might nevertheless freeze when faced with a new (*previously unseen*) question. By comparison, imagine another student, Inductive Irene, with comparably poor memorization skills, but a knack for picking up patterns. Note that if the exam truly consisted of recycled questions from a previous year, Ellie would handily outperform Irene. Even if Irene's inferred patterns yielded 90% accurate predictions, they could never compete with Ellie's 100% recall. However, even if the exam consisted entirely of fresh questions, Irene might maintain her 90% average.

As machine learning scientists, our goal is to discover *patterns*. But how can we be sure that we have truly discovered a *general* pattern and not simply memorized our data? Most of the time, our predictions are only useful if our model discovers such a pattern. We do not want to predict yesterday's stock prices, but tomorrow's. We do not need to recognize already diagnosed diseases for previously seen patients, but rather previously undiagnosed ailments in previously unseen patients. This problem—how to discover patterns that *generalize*—is the fundamental problem of machine learning, and arguably of all of statistics. We might cast this problem as just one slice of a far grander question that engulfs all of science: when are we ever justified in making the leap from particular observations to more general statements?

In real life, we must fit our models using a finite collection of data. The typical scales of that data vary wildly across domains. For many important medical problems, we can only access a few thousand data points. When studying rare diseases, we might be lucky to access hundreds. By contrast, the largest public datasets consisting of labeled photographs, e.g., ImageNet (Deng *et al.*, 2009), contain millions of images. And some unlabeled image collections such as the Flickr YFC100M dataset can be even larger, containing over 100 million images (Thomee *et al.*, 2016). However, even at this extreme scale, the number of available data points remains infinitesimally small compared to the space of all possible images at a megapixel resolution. Whenever we work with finite samples, we must keep in mind the risk that we might fit our training data, only to discover that we failed to discover a generalizable pattern.

The phenomenon of fitting closer to our training data than to the underlying distribution is called *overfitting*, and techniques for combatting overfitting are often called *regularization* methods. While it is no substitute for a proper introduction to statistical learning theory (see Boucheron *et al.* (2005), Vapnik (1998)), we will give you just enough intuition to get going. We will revisit generalization in many chapters throughout the book, exploring both what is known about the principles underlying generalization in various models, and also heuristic techniques that have been found (empirically) to yield improved generalization on tasks of practical interest.

### 3.6.1 Training Error and Generalization Error

In the standard supervised learning setting, we assume that the training data and the test data are drawn *independently* from *identical* distributions. This is commonly called the *IID assumption*. While this assumption is strong, it is worth noting that, absent any such assumption, we would be dead in the water. Why should we believe that training data sampled from distribution $P(X, Y)$ should tell us how to make predictions on test data generated by a *different distribution* $Q(X, Y)$? Making such leaps turns out to require strong assumptions about how $P$ and $Q$ are related. Later on we will discuss some assumptions that allow for shifts in distribution but first we need to understand the IID case, where $P(\cdot) = Q(\cdot)$.

To begin with, we need to differentiate between the *training error* $R_{\textrm{emp}}$, which is a *statistic* calculated on the training dataset, and the *generalization error* $R$, which is an *expectation* taken with respect to the underlying distribution. You can think of the generalization error as what you would see if you applied your model to an infinite stream of additional data examples drawn from the same underlying data distribution. Formally the training error is expressed as a *sum* (with the same notation as Section 3.1):

$$R_\textrm{emp}[\mathbf{X}, \mathbf{y}, f] = \frac{1}{n} \sum_{i=1}^n l(\mathbf{x}^{(i)}, y^{(i)}, f(\mathbf{x}^{(i)})), \tag{3.6.1}$$

while the generalization error is expressed as an integral:

$$R[p, f] = E_{(\mathbf{x}, y) \sim P} [l(\mathbf{x}, y, f(\mathbf{x}))] = \int \int l(\mathbf{x}, y, f(\mathbf{x})) p(\mathbf{x}, y) \ d\mathbf{x} dy. \tag{3.6.2}$$

Problematically, we can never calculate the generalization error $R$ exactly. Nobody ever tells us the precise form of the density function $p(\mathbf{x}, y)$. Moreover, we cannot sample an infinite stream of data points. Thus, in practice, we must *estimate* the generalization error by applying our model to an independent test set constituted of a random selection of examples $\mathbf{X}'$ and labels $\mathbf{y}'$ that were withheld from our training set. This consists of applying the same formula that was used for calculating the empirical training error but to a test set $\mathbf{X}', \mathbf{y}'$.

Crucially, when we evaluate our classifier on the test set, we are working with a *fixed* classifier (it does not depend on the sample of the test set), and thus estimating its error is simply the problem of mean estimation. However the same cannot be said for the training set. Note that the model we wind up with depends explicitly on the selection of the training set and thus the training error will in general be a biased estimate of the true error on the underlying population. The central question of generalization is then when should we expect our training error to be close to the population error (and thus the generalization error).

### Model Complexity

In classical theory, when we have simple models and abundant data, the training and generalization errors tend to be close. However, when we work with more complex models and/or fewer examples, we expect the training error to go down but the generalization gap to grow. This should not be surprising. Imagine a model class so expressive that for any dataset of $n$ examples, we can find a set of parameters that can perfectly fit arbitrary labels, even if randomly assigned. In this case, even if we fit our training data perfectly, how can we conclude anything about the generalization error? For all we know, our generalization error might be no better than random guessing.

In general, absent any restriction on our model class, we cannot conclude, based on fitting the training data alone, that our model has discovered any generalizable pattern (Vapnik *et al.*, 1994). On the other hand, if our model class was not capable of fitting arbitrary labels, then it must have discovered a pattern. Learning-theoretic ideas about model complexity derived some inspiration from the ideas of Karl Popper, an influential philosopher of science, who formalized the criterion of falsifiability. According to Popper, a theory that can explain any and all observations is not a scientific theory at all! After all, what has it told us about the world if it has not ruled out any possibility? In short, what we want is a hypothesis that *could not* explain any observations we might conceivably make and yet nevertheless happens to be compatible with those observations that we *in fact* make.

Now what precisely constitutes an appropriate notion of model complexity is a complex matter. Often, models with more parameters are able to fit a greater number of arbitrarily assigned labels. However, this is not necessarily true. For instance, kernel methods operate in spaces with infinite numbers of parameters, yet their complexity is controlled by other means (Schölkopf and Smola, 2002). One notion of complexity that often proves useful is the range of values that the parameters can take. Here, a model whose parameters are permitted to take arbitrary values would be more complex. We will revisit this idea in the next section, when we introduce *weight decay*, your first practical regularization technique. Notably, it can be difficult to compare complexity among members of substantially different model classes (say, decision trees vs. neural networks).

At this point, we must stress another important point that we will revisit when introducing deep neural networks. When a model is capable of fitting arbitrary labels, low training error does not necessarily imply low generalization error. *However, it does not necessarily imply high generalization error either!* All we can say with confidence is that low training error alone is not enough to certify low generalization error. Deep neural networks turn out to be just such models: while they generalize well in practice, they are too powerful to allow us to conclude much on the basis of training error alone. In these cases we must rely more heavily on our holdout data to certify generalization after the fact. Error on the holdout data, i.e., validation set, is called the *validation error*.

## 3.6.2 Underfitting or Overfitting?

When we compare the training and validation errors, we want to be mindful of two common situations. First, we want to watch out for cases when our training error and validation error are both substantial but there is a little gap between them. If the model is unable to reduce the training error, that could mean that our model is too simple (i.e., insufficiently expressive) to capture the pattern that we are trying to model. Moreover, since the *generalization gap* ($R_{\textrm{emp}} - R$) between our training and generalization errors is small, we have reason to believe that we could get away with a more complex model. This phenomenon is known as *underfitting*.

On the other hand, as we discussed above, we want to watch out for the cases when our training error is significantly lower than our validation error, indicating severe *overfitting*. Note that overfitting is not always a bad thing. In deep learning especially, the best predictive models often perform far better on training data than on holdout data. Ultimately, we usually care about driving the generalization error lower, and only care about the gap insofar as it becomes an obstacle to that end. Note that if the training error is zero, then the generalization gap is precisely equal to the generalization error and we can make progress only by reducing the gap.

### Polynomial Curve Fitting

To illustrate some classical intuition about overfitting and model complexity, consider the following: given training data consisting of a single feature $x$ and a corresponding real-valued label $y$, we try to find the polynomial of degree $d$

$$\hat{y}= \sum_{i=0}^d x^i w_i \tag{3.6.3}$$

for estimating the label $y$. This is just a linear regression problem where our features are given by the powers of $x$, the model's weights are given by $w_i$, and the bias is given by $w_0$ since $x^0 = 1$ for all $x$. Since this is just a linear regression problem, we can use the squared error as our loss function.

A higher-order polynomial function is more complex than a lower-order polynomial function, since the higher-order polynomial has more parameters and the model function's selection range is wider. Fixing the training dataset, higher-order polynomial functions should always achieve lower (at worst, equal) training error relative to lower-degree polynomials. In fact, whenever each data example has a distinct value of $x$, a polynomial function with degree equal to the number of data examples can fit the training set perfectly. We compare the relationship between polynomial degree (model complexity) and both underfitting and overfitting in Fig. 3.6.1.

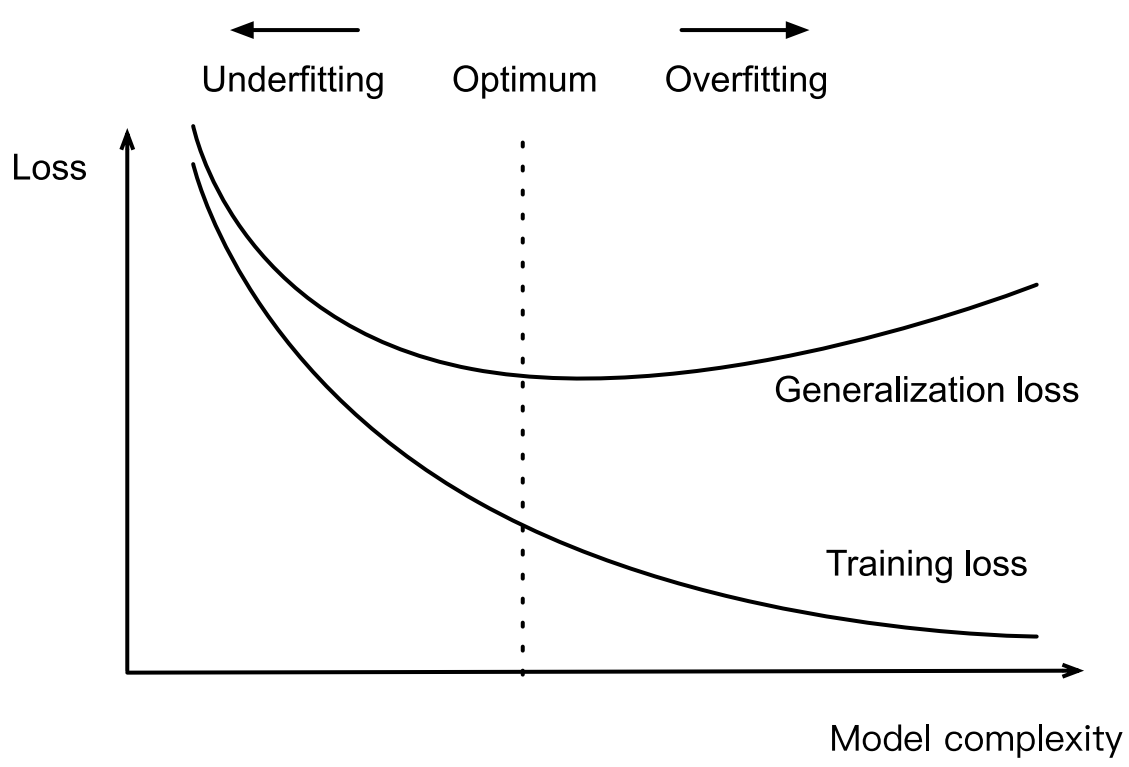

Fig. 3.6.1 Influence of model complexity on underfitting and overfitting.

#### Dataset Size

As the above bound already indicates, another big consideration to bear in mind is dataset size. Fixing our model, the fewer samples we have in the training dataset, the more likely (and more severely) we are to encounter overfitting. As we increase the amount of training data, the generalization error typically decreases. Moreover, in general, more data never hurts. For a fixed task and data distribution, model complexity should not increase more rapidly than the amount of data. Given more data, we might attempt to fit a more complex model. Absent sufficient data, simpler models may be more difficult to beat. For many tasks, deep learning only outperforms linear models when many thousands of training examples are available. In part, the current success of deep learning owes considerably to the abundance of massive datasets arising from Internet companies, cheap storage, connected devices, and the broad digitization of the economy.

### 3.6.3 Model Selection

Typically, we select our final model only after evaluating multiple models that differ in various ways (different architectures, training objectives, selected features, data preprocessing, learning rates, etc.). Choosing among many models is aptly called *model selection*.

In principle, we should not touch our test set until after we have chosen all our hyperparameters. Were we to use the test data in the model selection process, there is a risk that we might overfit the test data. Then we would be in serious trouble. If we overfit our training data, there is always the evaluation on test data to keep us honest. But if we overfit the test data, how would we ever know? See Ong *et al.* (2005) for an example of how this can lead to absurd results even for models where the complexity can be tightly controlled.

Thus, we should never rely on the test data for model selection. And yet we cannot rely solely on the training data for model selection either because we cannot estimate the generalization error on the very data that we use to train the model.

In practical applications, the picture gets muddier. While ideally we would only touch the test data once, to assess the very best model or to compare a small number of models with each other, real-world test data is seldom discarded after just one use. We can seldom afford a new test set for each round of experiments. In fact, recycling benchmark data for decades can have a significant impact on the development of algorithms, e.g., for image classification [81] and optical character recognition[82].

81 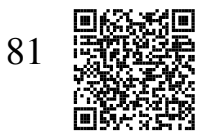

82

The common practice for addressing the problem of *training on the test set* is to split our data three ways, incorporating a *validation set* in addition to the training and test datasets. The result is a murky business where the boundaries between validation and test data are worryingly ambiguous. Unless explicitly stated otherwise, in the experiments in this book we are really working with what should rightly be called training data and validation data, with no true test sets. Therefore, the accuracy reported in each experiment of the book is really the validation accuracy and not a true test set accuracy.

#### Cross-Validation

When training data is scarce, we might not even be able to afford to hold out enough data to constitute a proper validation set. One popular solution to this problem is to employ *$K$-fold cross-validation*. Here, the original training data is split into $K$ non-overlapping subsets. Then model training and validation are executed $K$ times, each time training on $K-1$ subsets and validating on a different subset (the one not used for training in that round). Finally, the training and validation errors are estimated by averaging over the results from the $K$ experiments.

### 3.6.4 Summary

This section explored some of the underpinnings of generalization in machine learning. Some of these ideas become complicated and counterintuitive when we get to deeper models; here, models are capable of overfitting data badly, and the relevant notions of complexity can be both implicit and counterintuitive (e.g., larger architectures with more parameters generalizing better). We leave you with a few rules of thumb:

1. Use validation sets (or *$K$-fold cross-validation*) for model selection;
2. More complex models often require more data;
3. Relevant notions of complexity include both the number of parameters and the range of values that they are allowed to take;
4. Keeping all else equal, more data almost always leads to better generalization;
5. This entire talk of generalization is all predicated on the IID assumption. If we relax this assumption, allowing for distributions to shift between the train and testing periods, then we cannot say anything about generalization absent a further (perhaps milder) assumption.

### 3.6.5 Exercises

1. When can you solve the problem of polynomial regression exactly?
2. Give at least five examples where dependent random variables make treating the problem as IID data inadvisable.
3. Can you ever expect to see zero training error? Under which circumstances would you see zero generalization error?
4. Why is $K$-fold cross-validation very expensive to compute?
5. Why is the $K$-fold cross-validation error estimate biased?
6. The VC dimension is defined as the maximum number of points that can be classified with arbitrary labels $\{\pm 1\}$ by a function of a class of functions. Why might this not be a good idea for measuring how complex the class of functions is? Hint: consider the magnitude of the functions.
7. Your manager gives you a difficult dataset on which your current algorithm does not perform so well. How would you justify to him that you need more data? Hint: you cannot increase the data but you can decrease it.

Discussions[83].

## 3.7 Weight Decay

Now that we have characterized the problem of overfitting, we can introduce our first *regularization* technique. Recall that we can always mitigate overfitting by collecting more training data. However, that can be costly, time consuming, or entirely out of our control, making it impossible in the short run. For now, we can assume that we already have as much high-quality data as our resources permit and focus the tools at our disposal when the dataset is taken as a given.

Recall that in our polynomial regression example (Section 3.6.2) we could limit our model's capacity by tweaking the degree of the fitted polynomial. Indeed, limiting the number of features is a popular technique for mitigating overfitting. However, simply tossing aside features can be too blunt an instrument. Sticking with the polynomial regression example, consider what might happen with high-dimensional input. The natural extensions of polynomials to multivariate data are called *monomials*, which are simply products of powers of variables. The degree of a monomial is the sum of the powers. For example, $x_1^2 x_2$, and $x_3 x_5^2$ are both monomials of degree 3.

Note that the number of terms with degree $d$ blows up rapidly as $d$ grows larger. Given $k$ variables, the number of monomials of degree $d$ is $\binom{k-1+d}{k-1}$. Even small changes in degree, say from 2 to 3, dramatically increase the complexity of our model. Thus we often need a more fine-grained tool for adjusting function complexity.

```
%matplotlib inline
import torch
from torch import nn
from d2l import torch as d2l
```

## 3.7.1 Norms and Weight Decay

Rather than directly manipulating the number of parameters, *weight decay*, operates by restricting the values that the parameters can take. More commonly called $\ell_2$ regularization outside of deep learning circles when optimized by minibatch stochastic gradient descent, weight decay might be the most widely used technique for regularizing parametric machine learning models. The technique is motivated by the basic intuition that among all functions $f$, the function $f = 0$ (assigning the value 0 to all inputs) is in some sense the *simplest*, and that we can measure the complexity of a function by the distance of its parameters from zero. But how precisely should we measure the distance between a function and zero? There is no single right answer. In fact, entire branches of mathematics, including parts of functional analysis and the theory of Banach spaces, are devoted to addressing such issues.

One simple interpretation might be to measure the complexity of a linear function $f(\mathbf{x}) = \mathbf{w}^\top \mathbf{x}$ by some norm of its weight vector, e.g., $\|\mathbf{w}\|^2$. Recall that we introduced the $\ell_2$ norm and $\ell_1$ norm, which are special cases of the more general $\ell_p$ norm, in Section 2.3.11. The most common method for ensuring a small weight vector is to add its norm as a penalty term to the problem of minimizing the loss. Thus we replace our original objective, *minimizing the prediction loss on the training labels*, with new objective, *minimizing the sum of the prediction loss and the penalty term*. Now, if our weight vector grows too large, our learning algorithm might focus on minimizing the weight norm $\|\mathbf{w}\|^2$ rather than minimizing the training error. That is exactly what we want. To illustrate things in code, we revive our previous example from Section 3.1 for linear regression. There, our loss was given by

$$L(\mathbf{w}, b) = \frac{1}{n}\sum_{i=1}^n \frac{1}{2}\left(\mathbf{w}^\top \mathbf{x}^{(i)} + b - y^{(i)}\right)^2. \tag{3.7.1}$$

Recall that $\mathbf{x}^{(i)}$ are the features, $y^{(i)}$ is the label for any data example $i$, and $(\mathbf{w}, b)$ are the weight and bias parameters, respectively. To penalize the size of the weight vector, we must somehow add $\|\mathbf{w}\|^2$ to the loss function, but how should the model trade off the standard loss for this new additive penalty? In practice, we characterize this trade-off via the *regularization constant* $\lambda$, a nonnegative hyperparameter that we fit using validation data:

$$L(\mathbf{w}, b) + \frac{\lambda}{2} \|\mathbf{w}\|^2. \tag{3.7.2}$$

For $\lambda = 0$, we recover our original loss function. For $\lambda > 0$, we restrict the size of $\|\mathbf{w}\|$. We divide by 2 by convention: when we take the derivative of a quadratic function, the 2 and $1/2$ cancel out, ensuring that the expression for the update looks nice and simple. The astute reader might wonder why we work with the squared norm and not the standard norm (i.e., the Euclidean distance). We do this for computational convenience. By squaring the $\ell_2$ norm, we remove the square root, leaving the sum of squares of each component of the weight vector. This makes the derivative of the penalty easy to compute: the sum of derivatives equals the derivative of the sum.

Moreover, you might ask why we work with the $\ell_2$ norm in the first place and not, say, the $\ell_1$ norm. In fact, other choices are valid and popular throughout statistics. While $\ell_2$-regularized linear models constitute the classic *ridge regression* algorithm, $\ell_1$-regularized linear regression is a similarly fundamental method in statistics, popularly known as *lasso regression*. One reason to work with the $\ell_2$ norm is that it places an outsize penalty on large components of the weight vector. This biases our learning algorithm towards models that distribute weight evenly across a larger number of features. In practice, this might make them more robust to measurement error in a single variable. By contrast, $\ell_1$ penalties lead to models that concentrate weights on a small set of features by clearing the other weights to zero. This gives us an effective method for *feature selection*, which may be desirable for other reasons. For example, if our model only relies on a few features, then we may not need to collect, store, or transmit data for the other (dropped) features.

Using the same notation in (3.1.11), minibatch stochastic gradient descent updates for $\ell_2$-regularized regression as follows:

$$\mathbf{w} \leftarrow (1 - \eta\lambda)\, \mathbf{w} - \frac{\eta}{|\mathcal{B}|} \sum_{i \in \mathcal{B}} \mathbf{x}^{(i)} \left(\mathbf{w}^\top \mathbf{x}^{(i)} + b - y^{(i)}\right). \tag{3.7.3}$$

As before, we update $\mathbf{w}$ based on the amount by which our estimate differs from the observation. However, we also shrink the size of $\mathbf{w}$ towards zero. That is why the method is sometimes called "weight decay": given the penalty term alone, our optimization algorithm *decays* the weight at each step of training. In contrast to feature selection, weight decay offers us a mechanism for continuously adjusting the complexity of a function. Smaller values of $\lambda$ correspond to less constrained $\mathbf{w}$, whereas larger values of $\lambda$ constrain $\mathbf{w}$ more considerably. Whether we include a corresponding bias penalty $b^2$ can vary across implementations, and may vary across layers of a neural network. Often, we do not regularize the bias term. Besides, although $\ell_2$ regularization may not be equivalent to weight decay for other optimization algorithms, the idea of regularization through shrinking the size of weights still holds true.

### 3.7.2 High-Dimensional Linear Regression

We can illustrate the benefits of weight decay through a simple synthetic example.

First, we generate some data as before:

$$y = 0.05 + \sum_{i=1}^{d} 0.01 x_i + \epsilon \text{ where } \epsilon \sim \mathcal{N}(0, 0.01^2). \tag{3.7.4}$$

In this synthetic dataset, our label is given by an underlying linear function of our inputs,

corrupted by Gaussian noise with zero mean and standard deviation 0.01. For illustrative purposes, we can make the effects of overfitting pronounced, by increasing the dimensionality of our problem to $d = 200$ and working with a small training set with only 20 examples.

```
class Data(d2l.DataModule):
    def __init__(self, num_train, num_val, num_inputs, batch_size):
        self.save_hyperparameters()
        n = num_train + num_val
        self.X = torch.randn(n, num_inputs)
        noise = torch.randn(n, 1) * 0.01
        w, b = torch.ones((num_inputs, 1)) * 0.01, 0.05
        self.y = torch.matmul(self.X, w) + b + noise

    def get_dataloader(self, train):
        i = slice(0, self.num_train) if train else slice(self.num_train, None)
        return self.get_tensorloader([self.X, self.y], train, i)
```

## 3.7.3 Implementation from Scratch

Now, let's try implementing weight decay from scratch. Since minibatch stochastic gradient descent is our optimizer, we just need to add the squared $\ell_2$ penalty to the original loss function.

### Defining $\ell_2$ Norm Penalty

Perhaps the most convenient way of implementing this penalty is to square all terms in place and sum them.

```
def l2_penalty(w):
    return (w ** 2).sum() / 2
```

### Defining the Model

In the final model, the linear regression and the squared loss have not changed since Section 3.4, so we will just define a subclass of `d2l.LinearRegressionScratch`. The only change here is that our loss now includes the penalty term.

```
class WeightDecayScratch(d2l.LinearRegressionScratch):
    def __init__(self, num_inputs, lambd, lr, sigma=0.01):
        super().__init__(num_inputs, lr, sigma)
        self.save_hyperparameters()

    def loss(self, y_hat, y):
        return (super().loss(y_hat, y) +
                self.lambd * l2_penalty(self.w))
```

The following code fits our model on the training set with 20 examples and evaluates it on the validation set with 100 examples.

```
data = Data(num_train=20, num_val=100, num_inputs=200, batch_size=5)
trainer = d2l.Trainer(max_epochs=10)

def train_scratch(lambd):
    model = WeightDecayScratch(num_inputs=200, lambd=lambd, lr=0.01)
    model.board.yscale='log'
    trainer.fit(model, data)
    print('L2 norm of w:', float(l2_penalty(model.w)))
```

### Training without Regularization

We now run this code with `lambd = 0`, disabling weight decay. Note that we overfit badly, decreasing the training error but not the validation error—a textbook case of overfitting.

```
train_scratch(0)
```

```
L2 norm of w: 0.009889112785458565
```

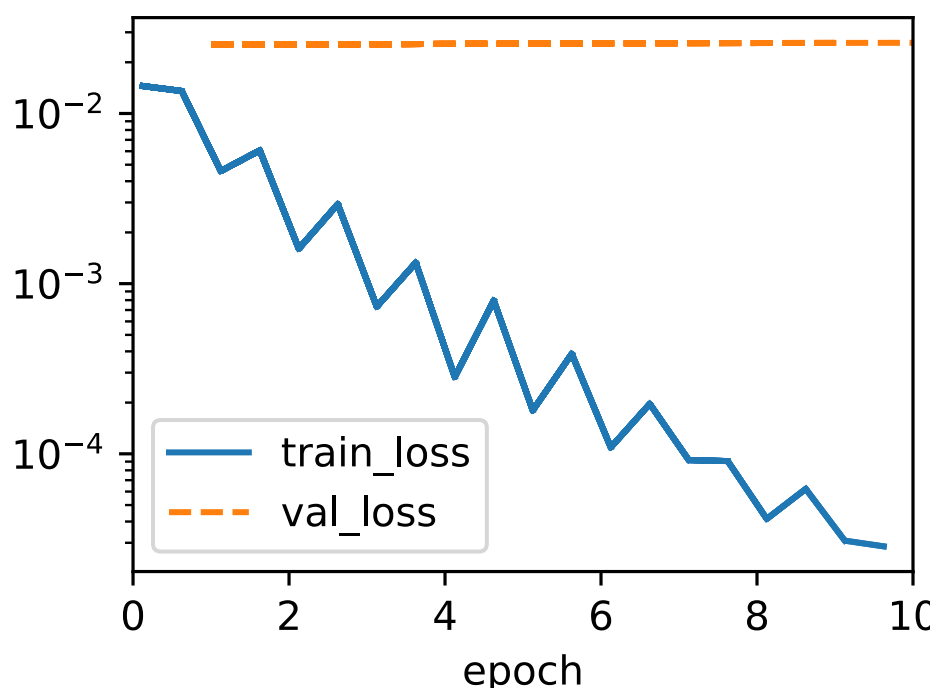

### Using Weight Decay

Below, we run with substantial weight decay. Note that the training error increases but the validation error decreases. This is precisely the effect we expect from regularization.

```
train_scratch(3)
```

```
L2 norm of w: 0.0014726519584655762
```

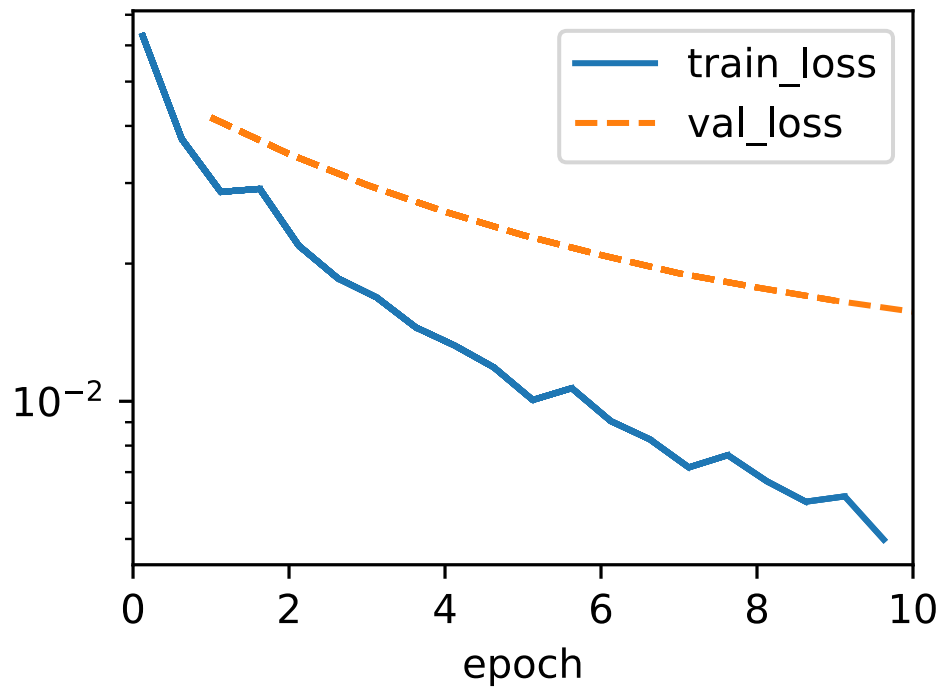

### 3.7.4 Concise Implementation

Because weight decay is ubiquitous in neural network optimization, the deep learning framework makes it especially convenient, integrating weight decay into the optimization algorithm itself for easy use in combination with any loss function. Moreover, this integration serves a computational benefit, allowing implementation tricks to add weight decay to the algorithm, without any additional computational overhead. Since the weight decay portion of the update depends only on the current value of each parameter, the optimizer must touch each parameter once anyway.

Below, we specify the weight decay hyperparameter directly through `weight_decay` when instantiating our optimizer. By default, PyTorch decays both weights and biases simultaneously, but we can configure the optimizer to handle different parameters according to different policies. Here, we only set `weight_decay` for the weights (the `net.weight` parameters), hence the bias (the `net.bias` parameter) will not decay.

```
class WeightDecay(d2l.LinearRegression):
    def __init__(self, wd, lr):
        super().__init__(lr)
        self.save_hyperparameters()
        self.wd = wd

    def configure_optimizers(self):
        return torch.optim.SGD([
            {'params': self.net.weight, 'weight_decay': self.wd},
            {'params': self.net.bias}], lr=self.lr)
```

The plot looks similar to that when we implemented weight decay from scratch. However, this version runs faster and is easier to implement, benefits that will become more pronounced as you address larger problems and this work becomes more routine.

```
model = WeightDecay(wd=3, lr=0.01)
model.board.yscale='log'
trainer.fit(model, data)

print('L2 norm of w:', float(l2_penalty(model.get_w_b()[0])))
```

```
L2 norm of w: 0.01231398992240429
```

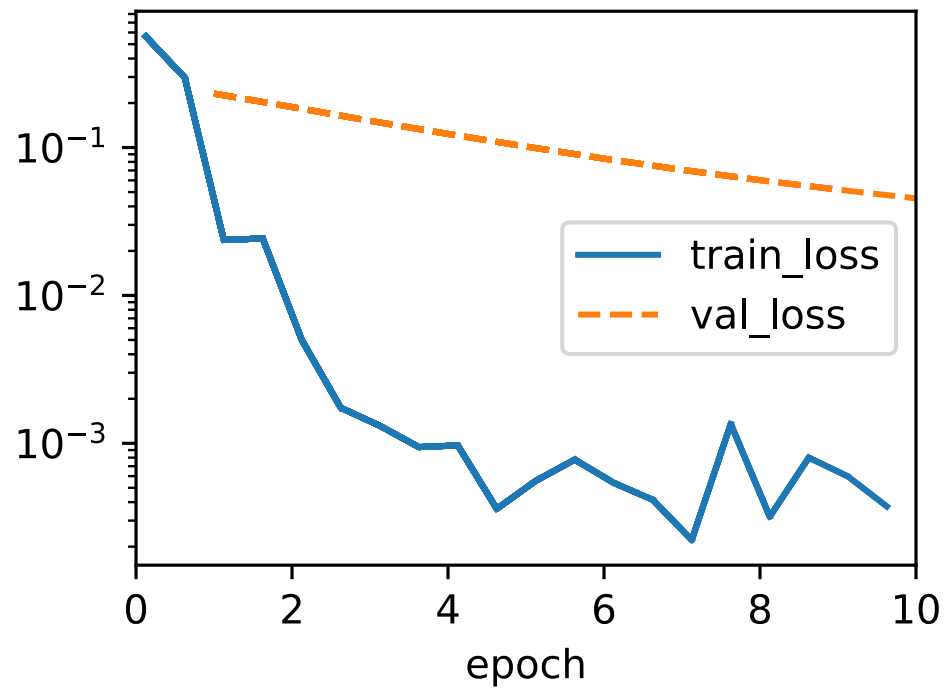

So far, we have touched upon one notion of what constitutes a simple linear function. However, even for simple nonlinear functions, the situation can be much more complex. To see this, the concept of reproducing kernel Hilbert space (RKHS)[84] allows one to apply tools introduced for linear functions in a nonlinear context. Unfortunately, RKHS-based algorithms tend to scale poorly to large, high-dimensional data. In this book we will often adopt the common heuristic whereby weight decay is applied to all layers of a deep network.

84

### 3.7.5 Summary

Regularization is a common method for dealing with overfitting. Classical regularization techniques add a penalty term to the loss function (when training) to reduce the complexity of the learned model. One particular choice for keeping the model simple is using an $\ell_2$ penalty. This leads to weight decay in the update steps of the minibatch stochastic gradient descent algorithm. In practice, the weight decay functionality is provided in optimizers from deep learning frameworks. Different sets of parameters can have different update behaviors within the same training loop.

### 3.7.6 Exercises

1. Experiment with the value of $\lambda$ in the estimation problem in this section. Plot training and validation accuracy as a function of $\lambda$. What do you observe?
2. Use a validation set to find the optimal value of $\lambda$. Is it really the optimal value? Does this matter?
3. What would the update equations look like if instead of $\|\mathbf{w}\|^2$ we used $\sum_i |w_i|$ as our penalty of choice ($\ell_1$ regularization)?
4. We know that $\|\mathbf{w}\|^2 = \mathbf{w}^\top \mathbf{w}$. Can you find a similar equation for matrices (see the Frobenius norm in Section 2.3.11)?

5. Review the relationship between training error and generalization error. In addition to weight decay, increased training, and the use of a model of suitable complexity, what other ways might help us deal with overfitting?
6. In Bayesian statistics we use the product of prior and likelihood to arrive at a posterior via $P(w \mid x) \propto P(x \mid w)P(w)$. How can you identify $P(w)$ with regularization?

Discussions[85].

85

# 4 Linear Neural Networks for Classification

Now that you have worked through all of the mechanics you are ready to apply the skills you have learned to broader kinds of tasks. Even as we pivot towards classification, most of the plumbing remains the same: loading the data, passing it through the model, generating output, calculating the loss, taking gradients with respect to weights, and updating the model. However, the precise form of the targets, the parametrization of the output layer, and the choice of loss function will adapt to suit the *classification* setting.

## 4.1 Softmax Regression

In Section 3.1, we introduced linear regression, working through implementations from scratch in Section 3.4 and again using high-level APIs of a deep learning framework in Section 3.5 to do the heavy lifting.

Regression is the hammer we reach for when we want to answer *how much?* or *how many?* questions. If you want to predict the number of dollars (price) at which a house will be sold, or the number of wins a baseball team might have, or the number of days that a patient will remain hospitalized before being discharged, then you are probably looking for a regression model. However, even within regression models, there are important distinctions. For instance, the price of a house will never be negative and changes might often be *relative* to its baseline price. As such, it might be more effective to regress on the logarithm of the price. Likewise, the number of days a patient spends in hospital is a *discrete nonnegative* random variable. As such, least mean squares might not be an ideal approach either. This sort of time-to-event modeling comes with a host of other complications that are dealt with in a specialized subfield called *survival modeling*.

The point here is not to overwhelm you but just to let you know that there is a lot more to estimation than simply minimizing squared errors. And more broadly, there is a lot more to supervised learning than regression. In this section, we focus on *classification* problems where we put aside *how much?* questions and instead focus on *which category?* questions.

- Does this email belong in the spam folder or the inbox?
- Is this customer more likely to sign up or not to sign up for a subscription service?

- Does this image depict a donkey, a dog, a cat, or a rooster?
- Which movie is Aston most likely to watch next?
- Which section of the book are you going to read next?

Colloquially, machine learning practitioners overload the word *classification* to describe two subtly different problems: (i) those where we are interested only in hard assignments of examples to categories (classes); and (ii) those where we wish to make soft assignments, i.e., to assess the probability that each category applies. The distinction tends to get blurred, in part, because often, even when we only care about hard assignments, we still use models that make soft assignments.

Even more, there are cases where more than one label might be true. For instance, a news article might simultaneously cover the topics of entertainment, business, and space flight, but not the topics of medicine or sports. Thus, categorizing it into one of the above categories on their own would not be very useful. This problem is commonly known as multi-label classification[86]. See Tsoumakas and Katakis (2007) for an overview and Huang *et al.* (2015) for an effective algorithm when tagging images.

86

## 4.1.1 Classification

To get our feet wet, let's start with a simple image classification problem. Here, each input consists of a $2 \times 2$ grayscale image. We can represent each pixel value with a single scalar, giving us four features $x_1, x_2, x_3, x_4$. Further, let's assume that each image belongs to one among the categories "cat", "chicken", and "dog".

Next, we have to choose how to represent the labels. We have two obvious choices. Perhaps the most natural impulse would be to choose $y \in \{1, 2, 3\}$, where the integers represent $\{\text{dog, cat, chicken}\}$ respectively. This is a great way of *storing* such information on a computer. If the categories had some natural ordering among them, say if we were trying to predict $\{\text{baby, toddler, adolescent, young adult, adult, geriatric}\}$, then it might even make sense to cast this as an ordinal regression[87] problem and keep the labels in this format. See Moon *et al.* (2010) for an overview of different types of ranking loss functions and Beutel *et al.* (2014) for a Bayesian approach that addresses responses with more than one mode.

In general, classification problems do not come with natural orderings among the classes. Fortunately, statisticians long ago invented a simple way to represent categorical data: the *one-hot encoding*. A one-hot encoding is a vector with as many components as we have categories. The component corresponding to a particular instance's category is set to 1 and all other components are set to 0. In our case, a label $y$ would be a three-dimensional vector, with $(1, 0, 0)$ corresponding to "cat", $(0, 1, 0)$ to "chicken", and $(0, 0, 1)$ to "dog":

$$y \in \{(1, 0, 0), (0, 1, 0), (0, 0, 1)\}. \tag{4.1.1}$$

### Linear Model

In order to estimate the conditional probabilities associated with all the possible classes, we need a model with multiple outputs, one per class. To address classification with linear models, we will need as many affine functions as we have outputs. Strictly speaking, we only need one fewer, since the final category has to be the difference between 1 and the sum of the other categories, but for reasons of symmetry we use a slightly redundant parametrization. Each output corresponds to its own affine function. In our case, since we have 4 features and 3 possible output categories, we need 12 scalars to represent the weights ($w$ with subscripts), and 3 scalars to represent the biases ($b$ with subscripts). This yields:

$$\begin{aligned} o_1 &= x_1 w_{11} + x_2 w_{12} + x_3 w_{13} + x_4 w_{14} + b_1,\\ o_2 &= x_1 w_{21} + x_2 w_{22} + x_3 w_{23} + x_4 w_{24} + b_2,\\ o_3 &= x_1 w_{31} + x_2 w_{32} + x_3 w_{33} + x_4 w_{34} + b_3. \end{aligned} \tag{4.1.2}$$

The corresponding neural network diagram is shown in Fig. 4.1.1. Just as in linear regression, we use a single-layer neural network. And since the calculation of each output, $o_1, o_2$, and $o_3$, depends on every input, $x_1$, $x_2$, $x_3$, and $x_4$, the output layer can also be described as a *fully connected layer*.

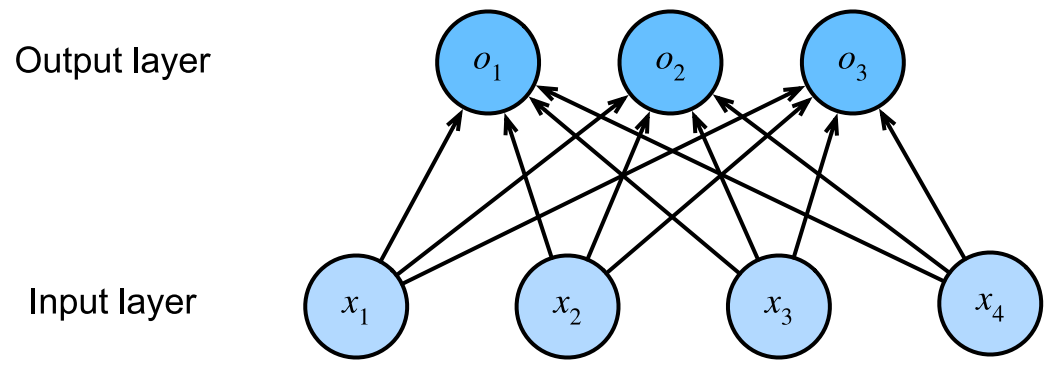

Fig. 4.1.1 Softmax regression is a single-layer neural network.

For a more concise notation we use vectors and matrices: $\mathbf{o} = \mathbf{W}\mathbf{x} + \mathbf{b}$ is much better suited for mathematics and code. Note that we have gathered all of our weights into a $3 \times 4$ matrix and all biases $\mathbf{b} \in \mathbb{R}^3$ in a vector.

### The Softmax

Assuming a suitable loss function, we could try, directly, to minimize the difference between $\mathbf{o}$ and the labels $\mathbf{y}$. While it turns out that treating classification as a vector-valued regression problem works surprisingly well, it is nonetheless unsatisfactory in the following ways:

- There is no guarantee that the outputs $o_i$ sum up to 1 in the way we expect probabilities to behave.
- There is no guarantee that the outputs $o_i$ are even nonnegative, even if their outputs sum up to 1, or that they do not exceed 1.

Both aspects render the estimation problem difficult to solve and the solution very brittle to outliers. For instance, if we assume that there is a positive linear dependency between the number of bedrooms and the likelihood that someone will buy a house, the probability might exceed 1 when it comes to buying a mansion! As such, we need a mechanism to "squish" the outputs.

There are many ways we might accomplish this goal. For instance, we could assume that the outputs $\mathbf{o}$ are corrupted versions of $\mathbf{y}$, where the corruption occurs by means of adding noise $\boldsymbol{\epsilon}$ drawn from a normal distribution. In other words, $\mathbf{y} = \mathbf{o} + \boldsymbol{\epsilon}$, where $\epsilon_i \sim \mathcal{N}(0, \sigma^2)$. This is the so-called probit model[88], first introduced by Fechner (1860). While appealing, it does not work quite as well nor lead to a particularly nice optimization problem, when compared to the softmax.

88 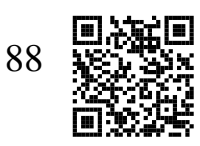

Another way to accomplish this goal (and to ensure nonnegativity) is to use an exponential function $P(y = i) \propto \exp o_i$. This does indeed satisfy the requirement that the conditional class probability increases with increasing $o_i$, it is monotonic, and all probabilities are nonnegative. We can then transform these values so that they add up to 1 by dividing each by their sum. This process is called *normalization*. Putting these two pieces together gives us the *softmax* function:

$$\hat{\mathbf{y}} = \mathrm{softmax}(\mathbf{o}) \quad \text{where} \quad \hat{y}_i = \frac{\exp(o_i)}{\sum_j \exp(o_j)}. \tag{4.1.3}$$

Note that the largest coordinate of $\mathbf{o}$ corresponds to the most likely class according to $\hat{\mathbf{y}}$. Moreover, because the softmax operation preserves the ordering among its arguments, we do not need to compute the softmax to determine which class has been assigned the highest probability. Thus,

$$\operatorname*{argmax}_j \hat{y}_j = \operatorname*{argmax}_j o_j. \tag{4.1.4}$$

The idea of a softmax dates back to Gibbs (1902), who adapted ideas from physics. Dating even further back, Boltzmann, the father of modern statistical physics, used this trick to model a distribution over energy states in gas molecules. In particular, he discovered that the prevalence of a state of energy in a thermodynamic ensemble, such as the molecules in a gas, is proportional to $\exp(-E/kT)$. Here, $E$ is the energy of a state, $T$ is the temperature, and $k$ is the Boltzmann constant. When statisticians talk about increasing or decreasing the "temperature" of a statistical system, they refer to changing $T$ in order to favor lower or higher energy states. Following Gibbs' idea, energy equates to error. Energy-based models (Ranzato *et al.*, 2007) use this point of view when describing problems in deep learning.

### Vectorization

To improve computational efficiency, we vectorize calculations in minibatches of data. Assume that we are given a minibatch $\mathbf{X} \in \mathbb{R}^{n \times d}$ of $n$ examples with dimensionality (number of inputs) $d$. Moreover, assume that we have $q$ categories in the output. Then the weights satisfy $\mathbf{W} \in \mathbb{R}^{d \times q}$ and the bias satisfies $\mathbf{b} \in \mathbb{R}^{1 \times q}$.

$$\begin{aligned} \mathbf{O} &= \mathbf{X}\mathbf{W} + \mathbf{b}, \\ \hat{\mathbf{Y}} &= \mathrm{softmax}(\mathbf{O}). \end{aligned} \tag{4.1.5}$$

This accelerates the dominant operation into a matrix–matrix product $\mathbf{X}\mathbf{W}$. Moreover, since each row in $\mathbf{X}$ represents a data example, the softmax operation itself can be computed *rowwise*: for each row of $\mathbf{O}$, exponentiate all entries and then normalize them by the sum. Note, though, that care must be taken to avoid exponentiating and taking logarithms of large numbers, since this can cause numerical overflow or underflow. Deep learning frameworks take care of this automatically.

## 4.1.2 Loss Function

Now that we have a mapping from features $\mathbf{x}$ to probabilities $\hat{\mathbf{y}}$, we need a way to optimize the accuracy of this mapping. We will rely on maximum likelihood estimation, the very same method that we encountered when providing a probabilistic justification for the mean squared error loss in Section 3.1.3.

### Log-Likelihood

The softmax function gives us a vector $\hat{\mathbf{y}}$, which we can interpret as the (estimated) conditional probabilities of each class, given any input $\mathbf{x}$, such as $\hat{y}_1 = P(y = \text{cat} \mid \mathbf{x})$. In the following we assume that for a dataset with features $\mathbf{X}$ the labels $\mathbf{Y}$ are represented using a one-hot encoding label vector. We can compare the estimates with reality by checking how probable the actual classes are according to our model, given the features:

$$P(\mathbf{Y} \mid \mathbf{X}) = \prod_{i=1}^{n} P(\mathbf{y}^{(i)} \mid \mathbf{x}^{(i)}). \tag{4.1.6}$$

We are allowed to use the factorization since we assume that each label is drawn independently from its respective distribution $P(\mathbf{y} \mid \mathbf{x}^{(i)})$. Since maximizing the product of terms is awkward, we take the negative logarithm to obtain the equivalent problem of minimizing the negative log-likelihood:

$$-\log P(\mathbf{Y} \mid \mathbf{X}) = \sum_{i=1}^{n} -\log P(\mathbf{y}^{(i)} \mid \mathbf{x}^{(i)}) = \sum_{i=1}^{n} l(\mathbf{y}^{(i)}, \hat{\mathbf{y}}^{(i)}), \tag{4.1.7}$$

where for any pair of label $\mathbf{y}$ and model prediction $\hat{\mathbf{y}}$ over $q$ classes, the loss function $l$ is

$$l(\mathbf{y}, \hat{\mathbf{y}}) = -\sum_{j=1}^{q} y_j \log \hat{y}_j. \tag{4.1.8}$$

For reasons explained later on, the loss function in (4.1.8) is commonly called the *cross-entropy loss*. Since $\mathbf{y}$ is a one-hot vector of length $q$, the sum over all its coordinates $j$ vanishes for all but one term. Note that the loss $l(\mathbf{y}, \hat{\mathbf{y}})$ is bounded from below by 0 whenever $\hat{\mathbf{y}}$ is a probability vector: no single entry is larger than 1, hence their negative logarithm cannot be lower than 0; $l(\mathbf{y}, \hat{\mathbf{y}}) = 0$ only if we predict the actual label with *certainty*. This can never happen for any finite setting of the weights because taking a softmax output towards 1 requires taking the corresponding input $o_i$ to infinity (or all other outputs $o_j$ for $j \neq i$ to negative infinity). Even if our model could assign an output probability of 0, any error made when assigning such high confidence would incur infinite loss ($-\log 0 = \infty$).

### Softmax and Cross-Entropy Loss

Since the softmax function and the corresponding cross-entropy loss are so common, it is worth understanding a bit better how they are computed. Plugging (4.1.3) into the definition of the loss in (4.1.8) and using the definition of the softmax we obtain

$$\begin{aligned} l(\mathbf{y}, \hat{\mathbf{y}}) &= - \sum_{j=1}^q y_j \log \frac{\exp(o_j)}{\sum_{k=1}^q \exp(o_k)} \\ &= \sum_{j=1}^q y_j \log \sum_{k=1}^q \exp(o_k) - \sum_{j=1}^q y_j o_j \\ &= \log \sum_{k=1}^q \exp(o_k) - \sum_{j=1}^q y_j o_j. \end{aligned} \tag{4.1.9}$$

To understand a bit better what is going on, consider the derivative with respect to any logit $o_j$. We get

$$\partial_{o_j} l(\mathbf{y}, \hat{\mathbf{y}}) = \frac{\exp(o_j)}{\sum_{k=1}^q \exp(o_k)} - y_j = \mathrm{softmax}(\mathbf{o})_j - y_j. \tag{4.1.10}$$

In other words, the derivative is the difference between the probability assigned by our model, as expressed by the softmax operation, and what actually happened, as expressed by elements in the one-hot label vector. In this sense, it is very similar to what we saw in regression, where the gradient was the difference between the observation $y$ and estimate $\hat{y}$. This is not a coincidence. In any exponential family model, the gradients of the log-likelihood are given by precisely this term. This fact makes computing gradients easy in practice.

Now consider the case where we observe not just a single outcome but an entire distribution over outcomes. We can use the same representation as before for the label $\mathbf{y}$. The only difference is that rather than a vector containing only binary entries, say $(0, 0, 1)$, we now have a generic probability vector, say $(0.1, 0.2, 0.7)$. The math that we used previously to define the loss $l$ in (4.1.8) still works well, just that the interpretation is slightly more general. It is the expected value of the loss for a distribution over labels. This loss is called the *cross-entropy loss* and it is one of the most commonly used losses for classification problems. We can demystify the name by introducing just the basics of information theory. In a nutshell, it measures the number of bits needed to encode what we see, $\mathbf{y}$, relative to what we predict that should happen, $\hat{\mathbf{y}}$. We provide a very basic explanation in the following. For further details on information theory see Cover and Thomas (1999) or MacKay (2003).

## 4.1.3 Information Theory Basics

Many deep learning papers use intuition and terms from information theory. To make sense of them, we need some common language. This is a survival guide. *Information theory* deals with the problem of encoding, decoding, transmitting, and manipulating information (also known as data).

### Entropy

The central idea in information theory is to quantify the amount of information contained in data. This places a limit on our ability to compress data. For a distribution $P$ its *entropy*, $H[P]$, is defined as:

$$H[P] = \sum_j -P(j) \log P(j). \tag{4.1.11}$$

One of the fundamental theorems of information theory states that in order to encode data drawn randomly from the distribution $P$, we need at least $H[P]$ "nats" to encode it (Shannon, 1948). If you wonder what a "nat" is, it is the equivalent of bit but when using a code with base $e$ rather than one with base 2. Thus, one nat is $\frac{1}{\log(2)} \approx 1.44$ bit.

### Surprisal

You might be wondering what compression has to do with prediction. Imagine that we have a stream of data that we want to compress. If it is always easy for us to predict the next token, then this data is easy to compress. Take the extreme example where every token in the stream always takes the same value. That is a very boring data stream! And not only it is boring, but it is also easy to predict. Because the tokens are always the same, we do not have to transmit any information to communicate the contents of the stream. Easy to predict, easy to compress.

However if we cannot perfectly predict every event, then we might sometimes be surprised. Our surprise is greater when an event is assigned lower probability. Claude Shannon settled on $\log \frac{1}{P(j)} = -\log P(j)$ to quantify one's *surprisal* at observing an event $j$ having assigned it a (subjective) probability $P(j)$. The entropy defined in (4.1.11) is then the *expected surprisal* when one assigned the correct probabilities that truly match the data-generating process.

### Cross-Entropy Revisited

So if entropy is the level of surprise experienced by someone who knows the true probability, then you might be wondering, what is cross-entropy? The cross-entropy *from P to Q*, denoted $H(P, Q)$, is the expected surprisal of an observer with subjective probabilities $Q$ upon seeing data that was actually generated according to probabilities $P$. This is given by $H(P, Q) \stackrel{\text{def}}{=}$ $\sum_j -P(j) \log Q(j)$. The lowest possible cross-entropy is achieved when $P = Q$. In this case, the cross-entropy from $P$ to $Q$ is $H(P, P) = H(P)$.

In short, we can think of the cross-entropy classification objective in two ways: (i) as maximizing the likelihood of the observed data; and (ii) as minimizing our surprisal (and thus the number of bits) required to communicate the labels.

### 4.1.4 Summary and Discussion

In this section, we encountered the first nontrivial loss function, allowing us to optimize over *discrete* output spaces. Key in its design was that we took a probabilistic approach, treating discrete categories as instances of draws from a probability distribution. As a side effect, we encountered the softmax, a convenient activation function that transforms outputs of an ordinary neural network layer into valid discrete probability distributions. We saw that the derivative of the cross-entropy loss when combined with softmax behaves very similarly to the derivative of squared error; namely by taking the difference between the expected behavior and its prediction. And, while we were only able to scratch the very surface of it, we encountered exciting connections to statistical physics and information theory.

While this is enough to get you on your way, and hopefully enough to whet your appetite, we hardly dived deep here. Among other things, we skipped over computational considerations. Specifically, for any fully connected layer with $d$ inputs and $q$ outputs, the parametrization and computational cost is $O(dq)$, which can be prohibitively high in practice. Fortunately, this cost of transforming $d$ inputs into $q$ outputs can be reduced through approximation and compression. For instance Deep Fried Convnets (Yang *et al.*, 2015) uses a combination of permutations, Fourier transforms, and scaling to reduce the cost from quadratic to log-linear. Similar techniques work for more advanced structural matrix approximations (Sindhwani *et al.*, 2015). Lastly, we can use quaternion-like decompositions to reduce the cost to $O(\frac{dq}{n})$, again if we are willing to trade off a small amount of accuracy for computational and storage cost (Zhang *et al.*, 2021) based on a compression factor $n$. This is an active area of research. What makes it challenging is that we do not necessarily strive for the most compact representation or the smallest number of floating point operations but rather for the solution that can be executed most efficiently on modern GPUs.

### 4.1.5 Exercises

1. We can explore the connection between exponential families and softmax in some more depth.
    1. Compute the second derivative of the cross-entropy loss $l(\mathbf{y}, \hat{\mathbf{y}})$ for softmax.
    2. Compute the variance of the distribution given by $\mathrm{softmax}(\mathbf{o})$ and show that it matches the second derivative computed above.
2. Assume that we have three classes which occur with equal probability, i.e., the probability vector is $(\frac{1}{3}, \frac{1}{3}, \frac{1}{3})$.

1. What is the problem if we try to design a binary code for it?
2. Can you design a better code? Hint: what happens if we try to encode two independent observations? What if we encode $n$ observations jointly?

3. When encoding signals transmitted over a physical wire, engineers do not always use binary codes. For instance, PAM-3[89] uses three signal levels $\{-1, 0, 1\}$ as opposed to two levels $\{0, 1\}$. How many ternary units do you need to transmit an integer in the range $\{0, \ldots, 7\}$? Why might this be a better idea in terms of electronics?

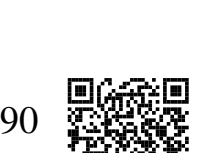

4. The Bradley–Terry model[90] uses a logistic model to capture preferences. For a user to choose between apples and oranges one assumes scores $o_{\mathrm{apple}}$ and $o_{\mathrm{orange}}$. Our requirements are that larger scores should lead to a higher likelihood in choosing the associated item and that the item with the largest score is the most likely one to be chosen (Bradley and Terry, 1952).

   1. Prove that softmax satisfies this requirement.
   2. What happens if you want to allow for a default option of choosing neither apples nor oranges? Hint: now the user has three choices.

5. Softmax gets its name from the following mapping: $\mathrm{RealSoftMax}(a, b) = \log(\exp(a) + \exp(b))$.

   1. Prove that $\mathrm{RealSoftMax}(a, b) > \max(a, b)$.
   2. How small can you make the difference between both functions? Hint: without loss of generality you can set $b = 0$ and $a \geq b$.
   3. Prove that this holds for $\lambda^{-1}\mathrm{RealSoftMax}(\lambda a, \lambda b)$, provided that $\lambda > 0$.
   4. Show that for $\lambda \to \infty$ we have $\lambda^{-1}\mathrm{RealSoftMax}(\lambda a, \lambda b) \to \max(a, b)$.
   5. Construct an analogous softmin function.
   6. Extend this to more than two numbers.

6. The function $g(\mathbf{x}) \stackrel{\mathrm{def}}{=} \log \sum_i \exp x_i$ is sometimes also referred to as the log-partition function[91].

   1. Prove that the function is convex. Hint: to do so, use the fact that the first derivative amounts to the probabilities from the softmax function and show that the second derivative is the variance.
   2. Show that $g$ is translation invariant, i.e., $g(\mathbf{x} + b) = g(\mathbf{x})$.
   3. What happens if some of the coordinates $x_i$ are very large? What happens if they're all very small?
   4. Show that if we choose $b = \max_i x_i$ we end up with a numerically stable implementation.

7. Assume that we have some probability distribution $P$. Suppose we pick another distribution $Q$ with $Q(i) \propto P(i)^\alpha$ for $\alpha > 0$.

1. Which choice of $\alpha$ corresponds to doubling the temperature? Which choice corresponds to halving it?
2. What happens if we let the temperature approach $0$?
3. What happens if we let the temperature approach $\infty$?

Discussions[92].

92

## 4.2 The Image Classification Dataset

93 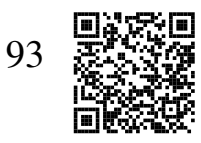

One widely used dataset for image classification is the MNIST dataset[93] (LeCun *et al.*, 1998) of handwritten digits. At the time of its release in the 1990s it posed a formidable challenge to most machine learning algorithms, consisting of 60,000 images of $28 \times 28$ pixels resolution (plus a test dataset of 10,000 images). To put things into perspective, back in 1995, a Sun SPARCStation 5 with a whopping 64MB of RAM and a blistering 5 MFLOPs was considered state of the art equipment for machine learning at AT&T Bell Laboratories. Achieving high accuracy on digit recognition was a key component in automating letter sorting for the USPS in the 1990s. Deep networks such as LeNet-5 (LeCun *et al.*, 1995), support vector machines with invariances (Schölkopf *et al.*, 1996), and tangent distance classifiers (Simard *et al.*, 1998) all could reach error rates below 1%.

For over a decade, MNIST served as *the* point of reference for comparing machine learning algorithms. While it had a good run as a benchmark dataset, even simple models by today's standards achieve classification accuracy over 95%, making it unsuitable for distinguishing between strong models and weaker ones. Even more, the dataset allows for *very* high levels of accuracy, not typically seen in many classification problems. This skewed algorithmic development towards specific families of algorithms that can take advantage of clean datasets, such as active set methods and boundary-seeking active set algorithms. Today, MNIST serves as more of a sanity check than as a benchmark. ImageNet (Deng *et al.*, 2009) poses a much more relevant challenge. Unfortunately, ImageNet is too large for many of the examples and illustrations in this book, as it would take too long to train to make the examples interactive. As a substitute we will focus our discussion in the coming sections on the qualitatively similar, but much smaller Fashion-MNIST dataset (Xiao *et al.*, 2017) which was released in 2017. It contains images of 10 categories of clothing at $28 \times 28$ pixels resolution.

```
%matplotlib inline
import time
import torch
import torchvision
from torchvision import transforms
from d2l import torch as d2l

d2l.use_svg_display()
```

### 4.2.1 Loading the Dataset

Since the Fashion-MNIST dataset is so useful, all major frameworks provide preprocessed versions of it. We can download and read it into memory using built-in framework utilities.

```
class FashionMNIST(d2l.DataModule):  #@save
    """The Fashion-MNIST dataset."""
    def __init__(self, batch_size=64, resize=(28, 28)):
        super().__init__()
        self.save_hyperparameters()
        trans = transforms.Compose([transforms.Resize(resize),
                                    transforms.ToTensor()])
        self.train = torchvision.datasets.FashionMNIST(
            root=self.root, train=True, transform=trans, download=True)
        self.val = torchvision.datasets.FashionMNIST(
            root=self.root, train=False, transform=trans, download=True)
```

Fashion-MNIST consists of images from 10 categories, each represented by 6000 images in the training dataset and by 1000 in the test dataset. A *test dataset* is used for evaluating model performance (it must not be used for training). Consequently the training set and the test set contain 60,000 and 10,000 images, respectively.

```
data = FashionMNIST(resize=(32, 32))
len(data.train), len(data.val)
```

```
(60000, 10000)
```

The images are grayscale and upscaled to $32 \times 32$ pixels in resolution above. This is similar to the original MNIST dataset which consisted of (binary) black and white images. Note, though, that most modern image data has three channels (red, green, blue) and that hyperspectral images can have in excess of 100 channels (the HyMap sensor has 126 channels). By convention we store an image as a $c \times h \times w$ tensor, where $c$ is the number of color channels, $h$ is the height and $w$ is the width.

```
data.train[0][0].shape
```

```
torch.Size([1, 32, 32])
```

The categories of Fashion-MNIST have human-understandable names. The following convenience method converts between numeric labels and their names.

```
@d2l.add_to_class(FashionMNIST)  #@save
def text_labels(self, indices):
    """Return text labels."""
    labels = ['t-shirt', 'trouser', 'pullover', 'dress', 'coat',
```

```
            'sandal', 'shirt', 'sneaker', 'bag', 'ankle boot']
    return [labels[int(i)] for i in indices]
```

### 4.2.2 Reading a Minibatch

To make our life easier when reading from the training and test sets, we use the built-in data iterator rather than creating one from scratch. Recall that at each iteration, a data iterator reads a minibatch of data with size `batch_size`. We also randomly shuffle the examples for the training data iterator.

```
@d2l.add_to_class(FashionMNIST)  #@save
def get_dataloader(self, train):
    data = self.train if train else self.val
    return torch.utils.data.DataLoader(data, self.batch_size, shuffle=train,
                                       num_workers=self.num_workers)
```

To see how this works, let's load a minibatch of images by invoking the `train_dataloader` method. It contains 64 images.

```
X, y = next(iter(data.train_dataloader()))
print(X.shape, X.dtype, y.shape, y.dtype)
```

```
torch.Size([64, 1, 32, 32]) torch.float32 torch.Size([64]) torch.int64
```

Let's look at the time it takes to read the images. Even though it is a built-in loader, it is not blazingly fast. Nonetheless, this is sufficient since processing images with a deep network takes quite a bit longer. Hence it is good enough that training a network will not be I/O constrained.

```
tic = time.time()
for X, y in data.train_dataloader():
    continue
f'{time.time() - tic:.2f} sec'
```

```
'5.41 sec'
```

### 4.2.3 Visualization

We will often be using the Fashion-MNIST dataset. A convenience function `show_images` can be used to visualize the images and the associated labels. Skipping implementation details, we just show the interface below: we only need to know how to invoke `d2l.show_images` rather than how it works for such utility functions.

```
def show_images(imgs, num_rows, num_cols, titles=None, scale=1.5):  #@save
    """Plot a list of images."""
    raise NotImplementedError
```

Let's put it to good use. In general, it is a good idea to visualize and inspect data that you are training on. Humans are very good at spotting oddities and because of that, visualization serves as an additional safeguard against mistakes and errors in the design of experiments. Here are the images and their corresponding labels (in text) for the first few examples in the training dataset.

```
@d2l.add_to_class(FashionMNIST)  #@save
def visualize(self, batch, nrows=1, ncols=8, labels=[]):
    X, y = batch
    if not labels:
        labels = self.text_labels(y)
    d2l.show_images(X.squeeze(1), nrows, ncols, titles=labels)
batch = next(iter(data.val_dataloader()))
data.visualize(batch)
```

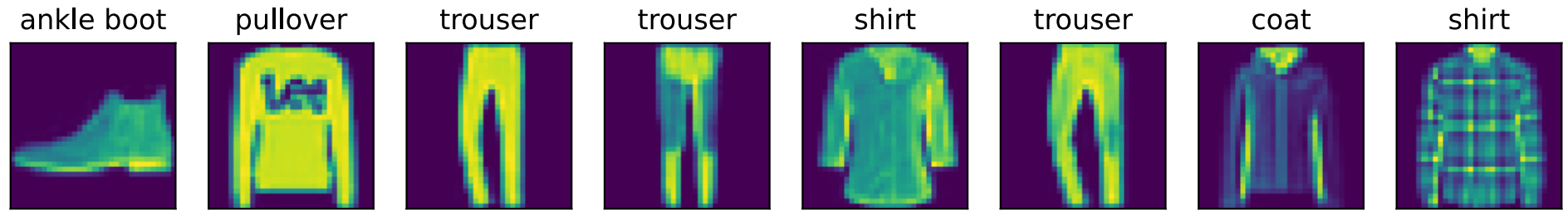

We are now ready to work with the Fashion-MNIST dataset in the sections that follow.

## 4.2.4 Summary

We now have a slightly more realistic dataset to use for classification. Fashion-MNIST is an apparel classification dataset consisting of images representing 10 categories. We will use this dataset in subsequent sections and chapters to evaluate various network designs, from a simple linear model to advanced residual networks. As we commonly do with images, we read them as a tensor of shape (batch size, number of channels, height, width). For now, we only have one channel as the images are grayscale (the visualization above uses a false color palette for improved visibility).

Lastly, data iterators are a key component for efficient performance. For instance, we might use GPUs for efficient image decompression, video transcoding, or other preprocessing. Whenever possible, you should rely on well-implemented data iterators that exploit high-performance computing to avoid slowing down your training loop.

## 4.2.5 Exercises

1. Does reducing the `batch_size` (for instance, to 1) affect the reading performance?

2. The data iterator performance is important. Do you think the current implementation is fast enough? Explore various options to improve it. Use a system profiler to find out where the bottlenecks are.

3. Check out the framework's online API documentation. Which other datasets are available?

Discussions[94].

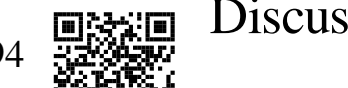

# 4.3 The Base Classification Model

You may have noticed that the implementations from scratch and the concise implementation using framework functionality were quite similar in the case of regression. The same is true for classification. Since many models in this book deal with classification, it is worth adding functionalities to support this setting specifically. This section provides a base class for classification models to simplify future code.

```
import torch
from d2l import torch as d2l
```

## 4.3.1 The `Classifier` Class

We define the `Classifier` class below. In the `validation_step` we report both the loss value and the classification accuracy on a validation batch. We draw an update for every `num_val_batches` batches. This has the benefit of generating the averaged loss and accuracy on the whole validation data. These average numbers are not exactly correct if the final batch contains fewer examples, but we ignore this minor difference to keep the code simple.

```
class Classifier(d2l.Module):  #@save
    """The base class of classification models."""
    def validation_step(self, batch):
        Y_hat = self(*batch[:-1])
        self.plot('loss', self.loss(Y_hat, batch[-1]), train=False)
        self.plot('acc', self.accuracy(Y_hat, batch[-1]), train=False)
```

By default we use a stochastic gradient descent optimizer, operating on minibatches, just as we did in the context of linear regression.

```
@d2l.add_to_class(d2l.Module)  #@save
def configure_optimizers(self):
    return torch.optim.SGD(self.parameters(), lr=self.lr)
```

### 4.3.2 Accuracy

Given the predicted probability distribution `y_hat`, we typically choose the class with the highest predicted probability whenever we must output a hard prediction. Indeed, many applications require that we make a choice. For instance, Gmail must categorize an email into "Primary", "Social", "Updates", "Forums", or "Spam". It might estimate probabilities internally, but at the end of the day it has to choose one among the classes.

When predictions are consistent with the label class `y`, they are correct. The classification accuracy is the fraction of all predictions that are correct. Although it can be difficult to optimize accuracy directly (it is not differentiable), it is often the performance measure that we care about the most. It is often *the* relevant quantity in benchmarks. As such, we will nearly always report it when training classifiers.

Accuracy is computed as follows. First, if `y_hat` is a matrix, we assume that the second dimension stores prediction scores for each class. We use `argmax` to obtain the predicted class by the index for the largest entry in each row. Then we compare the predicted class with the ground truth `y` elementwise. Since the equality operator `==` is sensitive to data types, we convert `y_hat`'s data type to match that of `y`. The result is a tensor containing entries of 0 (false) and 1 (true). Taking the sum yields the number of correct predictions.

```
@d2l.add_to_class(Classifier)  #@save
def accuracy(self, Y_hat, Y, averaged=True):
    """Compute the number of correct predictions."""
    Y_hat = Y_hat.reshape((-1, Y_hat.shape[-1]))
    preds = Y_hat.argmax(axis=1).type(Y.dtype)
    compare = (preds == Y.reshape(-1)).type(torch.float32)
    return compare.mean() if averaged else compare
```

### 4.3.3 Summary

Classification is a sufficiently common problem that it warrants its own convenience functions. Of central importance in classification is the *accuracy* of the classifier. Note that while we often care primarily about accuracy, we train classifiers to optimize a variety of other objectives for statistical and computational reasons. However, regardless of which loss function was minimized during training, it is useful to have a convenience method for assessing the accuracy of our classifier empirically.

### 4.3.4 Exercises

1. Denote by $L_\textrm{v}$ the validation loss, and let $L_\textrm{v}^\textrm{q}$ be its quick and dirty estimate computed by the loss function averaging in this section. Lastly, denote by $l_\textrm{v}^\textrm{b}$ the loss on the last minibatch. Express $L_\textrm{v}$ in terms of $L_\textrm{v}^\textrm{q}$, $l_\textrm{v}^\textrm{b}$, and the sample and minibatch sizes.
2. Show that the quick and dirty estimate $L_\textrm{v}^\textrm{q}$ is unbiased. That is, show that $E[L_\textrm{v}] = E[L_\textrm{v}^\textrm{q}]$. Why would you still want to use $L_\textrm{v}$ instead?

3. Given a multiclass classification loss, denoting by $l(y, y')$ the penalty of estimating $y'$ when we see $y$ and given a probabilty $p(y \mid x)$, formulate the rule for an optimal selection of $y'$. Hint: express the expected loss, using $l$ and $p(y \mid x)$.

Discussions[95].

95

## 4.4 Softmax Regression Implementation from Scratch

Because softmax regression is so fundamental, we believe that you ought to know how to implement it yourself. Here, we limit ourselves to defining the softmax-specific aspects of the model and reuse the other components from our linear regression section, including the training loop.

```
import torch
from d2l import torch as d2l
```

### 4.4.1 The Softmax

Let's begin with the most important part: the mapping from scalars to probabilities. For a refresher, recall the operation of the sum operator along specific dimensions in a tensor, as discussed in Section 2.3.6 and Section 2.3.7. Given a matrix `X` we can sum over all elements (by default) or only over elements in the same axis. The `axis` variable lets us compute row and column sums:

```
X = torch.tensor([[1.0, 2.0, 3.0], [4.0, 5.0, 6.0]])
X.sum(0, keepdims=True), X.sum(1, keepdims=True)
```

```
(tensor([[5., 7., 9.]]),
 tensor([[ 6.],
         [15.]]))
```

Computing the softmax requires three steps: (i) exponentiation of each term; (ii) a sum over each row to compute the normalization constant for each example; (iii) division of each row by its normalization constant, ensuring that the result sums to 1:

$$\mathrm{softmax}(\mathbf{X})_{ij} = \frac{\exp(\mathbf{X}_{ij})}{\sum_k \exp(\mathbf{X}_{ik})}. \tag{4.4.1}$$

The (logarithm of the) denominator is called the (log) *partition function*. It was introduced in statistical physics[96] to sum over all possible states in a thermodynamic ensemble. The implementation is straightforward:

96

```
def softmax(X):
    X_exp = torch.exp(X)
    partition = X_exp.sum(1, keepdims=True)
    return X_exp / partition  # The broadcasting mechanism is applied here
```

For any input X, we turn each element into a nonnegative number. Each row sums up to 1, as is required for a probability. Caution: the code above is *not* robust against very large or very small arguments. While it is sufficient to illustrate what is happening, you should *not* use this code verbatim for any serious purpose. Deep learning frameworks have such protections built in and we will be using the built-in softmax going forward.

```
X = torch.rand((2, 5))
X_prob = softmax(X)
X_prob, X_prob.sum(1)
```

```
(tensor([[0.1373, 0.2025, 0.2170, 0.2233, 0.2199],
         [0.2669, 0.1146, 0.2801, 0.1372, 0.2012]]),
 tensor([1., 1.]))
```

## 4.4.2 The Model

We now have everything that we need to implement the softmax regression model. As in our linear regression example, each instance will be represented by a fixed-length vector. Since the raw data here consists of $28 \times 28$ pixel images, we flatten each image, treating them as vectors of length 784. In later chapters, we will introduce convolutional neural networks, which exploit the spatial structure in a more satisfying way.

In softmax regression, the number of outputs from our network should be equal to the number of classes. Since our dataset has 10 classes, our network has an output dimension of 10. Consequently, our weights constitute a $784 \times 10$ matrix plus a $1 \times 10$ row vector for the biases. As with linear regression, we initialize the weights W with Gaussian noise. The biases are initialized as zeros.

```
class SoftmaxRegressionScratch(d2l.Classifier):
    def __init__(self, num_inputs, num_outputs, lr, sigma=0.01):
        super().__init__()
        self.save_hyperparameters()
        self.W = torch.normal(0, sigma, size=(num_inputs, num_outputs),
                              requires_grad=True)
        self.b = torch.zeros(num_outputs, requires_grad=True)

    def parameters(self):
        return [self.W, self.b]
```

The code below defines how the network maps each input to an output. Note that we flatten each $28 \times 28$ pixel image in the batch into a vector using reshape before passing the data through our model.

```
@d2l.add_to_class(SoftmaxRegressionScratch)
def forward(self, X):
    X = X.reshape((-1, self.W.shape[0]))
    return softmax(torch.matmul(X, self.W) + self.b)
```

### 4.4.3 The Cross-Entropy Loss

Next we need to implement the cross-entropy loss function (introduced in Section 4.1.2). This may be the most common loss function in all of deep learning. At the moment, applications of deep learning easily cast as classification problems far outnumber those better treated as regression problems.

Recall that cross-entropy takes the negative log-likelihood of the predicted probability assigned to the true label. For efficiency we avoid Python for-loops and use indexing instead. In particular, the one-hot encoding in $\mathbf{y}$ allows us to select the matching terms in $\hat{\mathbf{y}}$.

To see this in action we create sample data `y_hat` with 2 examples of predicted probabilities over 3 classes and their corresponding labels `y`. The correct labels are 0 and 2 respectively (i.e., the first and third class). Using `y` as the indices of the probabilities in `y_hat`, we can pick out terms efficiently.

```
y = torch.tensor([0, 2])
y_hat = torch.tensor([[0.1, 0.3, 0.6], [0.3, 0.2, 0.5]])
y_hat[[0, 1], y]
```

```
tensor([0.1000, 0.5000])
```

Now we can implement the cross-entropy loss function by averaging over the logarithms of the selected probabilities.

```
def cross_entropy(y_hat, y):
    return -torch.log(y_hat[list(range(len(y_hat))), y]).mean()

cross_entropy(y_hat, y)
```

```
tensor(1.4979)
```

```
@d2l.add_to_class(SoftmaxRegressionScratch)
def loss(self, y_hat, y):
    return cross_entropy(y_hat, y)
```

### 4.4.4 Training

We reuse the `fit` method defined in Section 3.4 to train the model with 10 epochs. Note that the number of epochs (`max_epochs`), the minibatch size (`batch_size`), and learning rate (`lr`) are adjustable hyperparameters. That means that while these values are not learned during our primary training loop, they still influence the performance of our model, both vis-à-vis training and generalization performance. In practice you will want to choose these values based on the *validation* split of the data and then, ultimately, to evaluate your final model on the *test* split. As discussed in Section 3.6.3, we will regard the test data of Fashion-MNIST as the validation set, thus reporting validation loss and validation accuracy on this split.

```
data = d2l.FashionMNIST(batch_size=256)
model = SoftmaxRegressionScratch(num_inputs=784, num_outputs=10, lr=0.1)
trainer = d2l.Trainer(max_epochs=10)
trainer.fit(model, data)
```

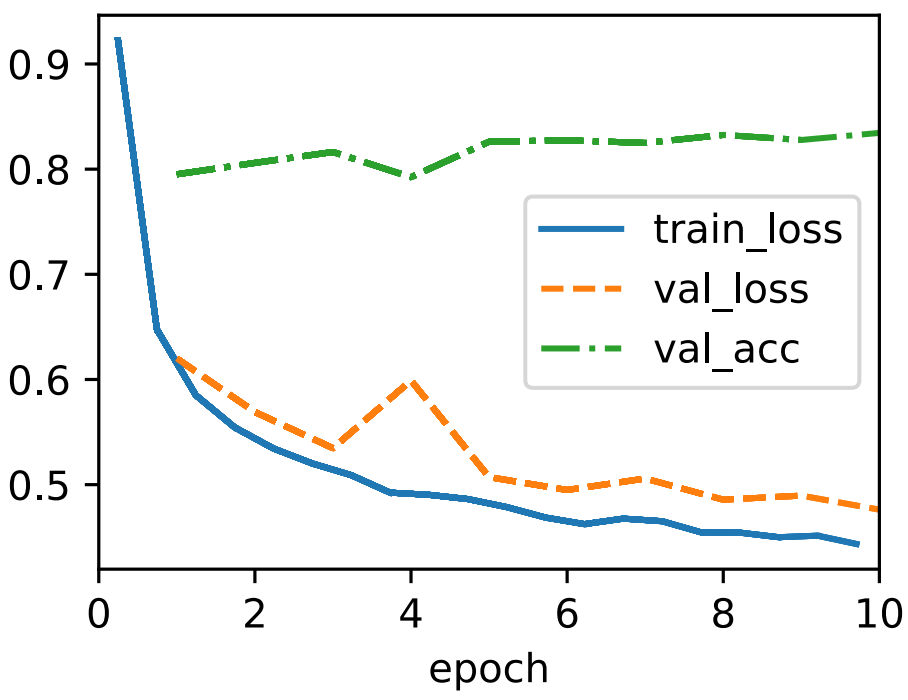

## 4.4.5 Prediction

Now that training is complete, our model is ready to classify some images.

```
X, y = next(iter(data.val_dataloader()))
preds = model(X).argmax(axis=1)
preds.shape
```

```
torch.Size([256])
```

We are more interested in the images we label *incorrectly*. We visualize them by comparing their actual labels (first line of text output) with the predictions from the model (second line of text output).

```
wrong = preds.type(y.dtype) != y
X, y, preds = X[wrong], y[wrong], preds[wrong]
```

```
labels = [a+'\n'+b for a, b in zip(
    data.text_labels(y), data.text_labels(preds))]
data.visualize([X, y], labels=labels)
```

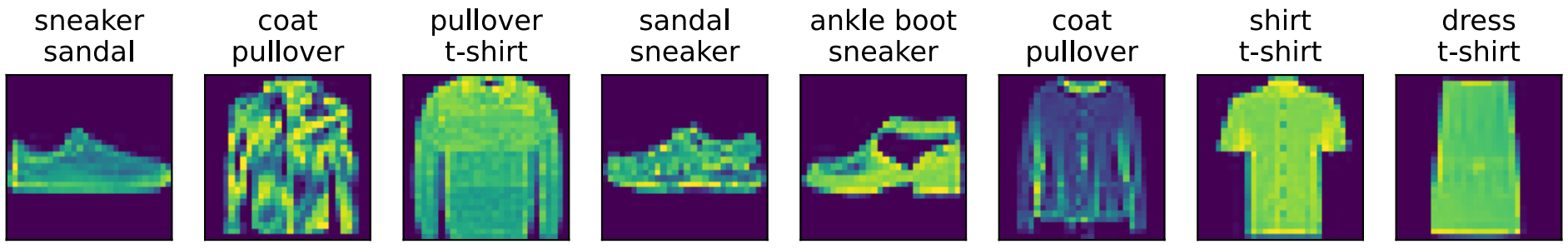

### 4.4.6 Summary

By now we are starting to get some experience with solving linear regression and classification problems. With it, we have reached what would arguably be the state of the art of 1960–1970s of statistical modeling. In the next section, we will show you how to leverage deep learning frameworks to implement this model much more efficiently.

### 4.4.7 Exercises

1. In this section, we directly implemented the softmax function based on the mathematical definition of the softmax operation. As discussed in Section 4.1 this can cause numerical instabilities.
    1. Test whether `softmax` still works correctly if an input has a value of $100$.
    2. Test whether `softmax` still works correctly if the largest of all inputs is smaller than $-100$?
    3. Implement a fix by looking at the value relative to the largest entry in the argument.
2. Implement a `cross_entropy` function that follows the definition of the cross-entropy loss function $\sum_i y_i \log \hat{y}_i$.
    1. Try it out in the code example of this section.
    2. Why do you think it runs more slowly?
    3. Should you use it? When would it make sense to?
    4. What do you need to be careful of? Hint: consider the domain of the logarithm.
3. Is it always a good idea to return the most likely label? For example, would you do this for medical diagnosis? How would you try to address this?
4. Assume that we want to use softmax regression to predict the next word based on some features. What are some problems that might arise from a large vocabulary?
5. Experiment with the hyperparameters of the code in this section. In particular:

1. Plot how the validation loss changes as you change the learning rate.
2. Do the validation and training loss change as you change the minibatch size? How large or small do you need to go before you see an effect?

Discussions[97].

97

# 4.5 Concise Implementation of Softmax Regression

Just as high-level deep learning frameworks made it easier to implement linear regression (see Section 3.5), they are similarly convenient here.

```
import torch
from torch import nn
from torch.nn import functional as F
from d2l import torch as d2l
```

## 4.5.1 Defining the Model

As in Section 3.5, we construct our fully connected layer using the built-in layer. The built-in `__call__` method then invokes `forward` whenever we need to apply the network to some input.

We use a `Flatten` layer to convert the fourth-order tensor `X` to second order by keeping the dimensionality along the first axis unchanged.

```
class SoftmaxRegression(d2l.Classifier):  #@save
    """The softmax regression model."""
    def __init__(self, num_outputs, lr):
        super().__init__()
        self.save_hyperparameters()
        self.net = nn.Sequential(nn.Flatten(),
                                 nn.LazyLinear(num_outputs))

    def forward(self, X):
        return self.net(X)
```

## 4.5.2 Softmax Revisited

In Section 4.4 we calculated our model's output and applied the cross-entropy loss. While this is perfectly reasonable mathematically, it is risky computationally, because of numerical underflow and overflow in the exponentiation.

Recall that the softmax function computes probabilities via $\hat{y}_j = \frac{\exp(o_j)}{\sum_k \exp(o_k)}$. If some of the $o_k$ are very large, i.e., very positive, then $\exp(o_k)$ might be larger than the largest number we can have for certain data types. This is called *overflow*. Likewise, if every argument is a very large negative number, we will get *underflow*. For instance, single precision floating point numbers approximately cover the range of $10^{-38}$ to $10^{38}$. As such, if the largest term in $\mathbf{o}$ lies outside the interval $[-90, 90]$, the result will not be stable. A way round this problem is to subtract $\bar{o} \stackrel{\mathrm{def}}{=} \max_k o_k$ from all entries:

$$\hat{y}_j = \frac{\exp o_j}{\sum_k \exp o_k} = \frac{\exp(o_j - \bar{o}) \exp \bar{o}}{\sum_k \exp(o_k - \bar{o}) \exp \bar{o}} = \frac{\exp(o_j - \bar{o})}{\sum_k \exp(o_k - \bar{o})}. \tag{4.5.1}$$

By construction we know that $o_j - \bar{o} \leq 0$ for all $j$. As such, for a $q$-class classification problem, the denominator is contained in the interval $[1, q]$. Moreover, the numerator never exceeds 1, thus preventing numerical overflow. Numerical underflow only occurs when $\exp(o_j - \bar{o})$ numerically evaluates as 0. Nonetheless, a few steps down the road we might find ourselves in trouble when we want to compute $\log \hat{y}_j$ as $\log 0$. In particular, in backpropagation, we might find ourselves faced with a screenful of the dreaded `NaN` (Not a Number) results.

Fortunately, we are saved by the fact that even though we are computing exponential functions, we ultimately intend to take their log (when calculating the cross-entropy loss). By combining softmax and cross-entropy, we can escape the numerical stability issues altogether. We have:

$$\log \hat{y}_j = \log \frac{\exp(o_j - \bar{o})}{\sum_k \exp(o_k - \bar{o})} = o_j - \bar{o} - \log \sum_k \exp(o_k - \bar{o}). \tag{4.5.2}$$

This avoids both overflow and underflow. We will want to keep the conventional softmax function handy in case we ever want to evaluate the output probabilities by our model. But instead of passing softmax probabilities into our new loss function, we just pass the logits and compute the softmax and its log all at once inside the cross-entropy loss function, which does smart things like the "LogSumExp trick"[98].

```python
@d2l.add_to_class(d2l.Classifier)  #@save
def loss(self, Y_hat, Y, averaged=True):
    Y_hat = Y_hat.reshape((-1, Y_hat.shape[-1]))
    Y = Y.reshape((-1,))
    return F.cross_entropy(
        Y_hat, Y, reduction='mean' if averaged else 'none')
```

### 4.5.3 Training

Next we train our model. We use Fashion-MNIST images, flattened to 784-dimensional feature vectors.

```python
data = d2l.FashionMNIST(batch_size=256)
model = SoftmaxRegression(num_outputs=10, lr=0.1)
trainer = d2l.Trainer(max_epochs=10)
trainer.fit(model, data)
```

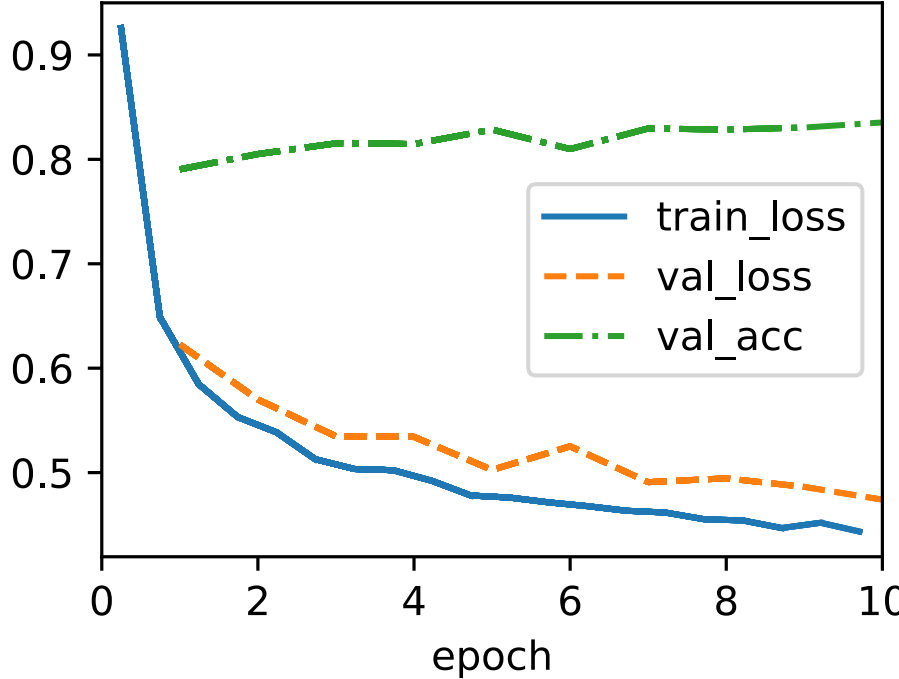

As before, this algorithm converges to a solution that is reasonably accurate, albeit this time with fewer lines of code than before.

### 4.5.4 Summary

High-level APIs are very convenient at hiding from their user potentially dangerous aspects, such as numerical stability. Moreover, they allow users to design models concisely with very few lines of code. This is both a blessing and a curse. The obvious benefit is that it makes things highly accessible, even to engineers who never took a single class of statistics in their life (in fact, they are part of the target audience of the book). But hiding the sharp edges also comes with a price: a disincentive to add new and different components on your own, since there is little muscle memory for doing it. Moreover, it makes it more difficult to *fix* things whenever the protective padding of a framework fails to cover all the corner cases entirely. Again, this is due to lack of familiarity.

As such, we strongly urge you to review *both* the bare bones and the elegant versions of many of the implementations that follow. While we emphasize ease of understanding, the implementations are nonetheless usually quite performant (convolutions are the big exception here). It is our intention to allow you to build on these when you invent something new that no framework can give you.

### 4.5.5 Exercises

1. Deep learning uses many different number formats, including FP64 double precision (used extremely rarely), FP32 single precision, BFLOAT16 (good for compressed representations), FP16 (very unstable), TF32 (a new format from NVIDIA), and INT8. Compute the smallest and largest argument of the exponential function for which the result does not lead to numerical underflow or overflow.
2. INT8 is a very limited format consisting of nonzero numbers from $1$ to $255$. How could you extend its dynamic range without using more bits? Do standard multiplication and addition still work?

3. Increase the number of epochs for training. Why might the validation accuracy decrease after a while? How could we fix this?
4. What happens as you increase the learning rate? Compare the loss curves for several learning rates. Which one works better? When?

Discussions[99].

99

## 4.6 Generalization in Classification

So far, we have focused on how to tackle multiclass classification problems by training (linear) neural networks with multiple outputs and softmax functions. Interpreting our model's outputs as probabilistic predictions, we motivated and derived the cross-entropy loss function, which calculates the negative log likelihood that our model (for a fixed set of parameters) assigns to the actual labels. And finally, we put these tools into practice by fitting our model to the training set. However, as always, our goal is to learn *general patterns*, as assessed empirically on previously unseen data (the test set). High accuracy on the training set means nothing. Whenever each of our inputs is unique (and indeed this is true for most high-dimensional datasets), we can attain perfect accuracy on the training set by just memorizing the dataset on the first training epoch, and subsequently looking up the label whenever we see a new image. And yet, memorizing the exact labels associated with the exact training examples does not tell us how to classify new examples. Absent further guidance, we might have to fall back on random guessing whenever we encounter new examples.

A number of burning questions demand immediate attention:

1. How many test examples do we need to give a good estimate of the accuracy of our classifiers on the underlying population?
2. What happens if we keep evaluating models on the same test repeatedly?
3. Why should we expect that fitting our linear models to the training set should fare any better than our naive memorization scheme?

Whereas Section 3.6 introduced the basics of overfitting and generalization in the context of linear regression, this chapter will go a little deeper, introducing some of the foundational ideas of statistical learning theory. It turns out that we often can guarantee generalization *a priori*: for many models, and for any desired upper bound on the generalization gap $\epsilon$, we can often determine some required number of samples $n$ such that if our training set contains at least $n$ samples, our empirical error will lie within $\epsilon$ of the true error, *for any data generating distribution*. Unfortunately, it also turns out that while these sorts of guarantees provide a profound set of intellectual building blocks, they are of limited practical utility to the deep learning practitioner. In short, these guarantees suggest that ensuring generalization of deep neural networks *a priori* requires an absurd number of examples (perhaps trillions or more), even when we find that, on the tasks we care about, deep neural networks typically generalize remarkably well with far fewer examples (thousands). Thus deep learning practitioners often forgo *a priori* guarantees altogether, instead employing methods that have generalized well on similar problems in the past, and certifying generalization *post hoc* through empirical evaluations. When we get to Chapter 5, we will revisit generalization and provide a light introduction to the vast scientific literature that has sprung in attempts to explain why deep neural networks generalize in practice.

### 4.6.1 The Test Set

Since we have already begun to rely on test sets as the gold standard method for assessing generalization error, let's get started by discussing the properties of such error estimates. Let's focus on a fixed classifier $f$, without worrying about how it was obtained. Moreover suppose that we possess a *fresh* dataset of examples $\mathcal{D} = (\mathbf{x}^{(i)}, y^{(i)})_{i=1}^n$ that were not used to train the classifier $f$. The *empirical error* of our classifier $f$ on $\mathcal{D}$ is simply the fraction of instances for which the prediction $f(\mathbf{x}^{(i)})$ disagrees with the true label $y^{(i)}$, and is given by the following expression:

$$\epsilon_\mathcal{D}(f) = \frac{1}{n}\sum_{i=1}^n \mathbf{1}(f(\mathbf{x}^{(i)}) \neq y^{(i)}). \tag{4.6.1}$$

By contrast, the *population error* is the *expected* fraction of examples in the underlying population (some distribution $P(X,Y)$ characterized by probability density function $p(\mathbf{x},y)$) for which our classifier disagrees with the true label:

$$\epsilon(f) = E_{(\mathbf{x}, y) \sim P} \mathbf{1}(f(\mathbf{x}) \neq y) = \int\int \mathbf{1}(f(\mathbf{x}) \neq y) p(\mathbf{x}, y) \;d\mathbf{x} dy. \tag{4.6.2}$$

While $\epsilon(f)$ is the quantity that we actually care about, we cannot observe it directly, just as we cannot directly observe the average height in a large population without measuring every single person. We can only estimate this quantity based on samples. Because our test set $\mathcal{D}$ is statistically representative of the underlying population, we can view $\epsilon_\mathcal{D}(f)$ as a statistical estimator of the population error $\epsilon(f)$. Moreover, because our quantity of interest $\epsilon(f)$ is an expectation (of the random variable $\mathbf{1}(f(X) \neq Y)$) and the corresponding estimator $\epsilon_\mathcal{D}(f)$ is the sample average, estimating the population error is simply the classic problem of mean estimation, which you may recall from Section 2.6.

An important classical result from probability theory called the *central limit theorem* guarantees that whenever we possess $n$ random samples $a_1, \ldots, a_n$ drawn from any distribution with mean $\mu$ and standard deviation $\sigma$, then, as the number of samples $n$ approaches infinity, the sample average $\hat{\mu}$ approximately tends towards a normal distribution centered at the true mean and with standard deviation $\sigma/\sqrt{n}$. Already, this tells us something important: as the number of examples grows large, our test error $\epsilon_\mathcal{D}(f)$ should approach the true error $\epsilon(f)$ at a rate of $\mathcal{O}(1/\sqrt{n})$. Thus, to estimate our test error twice as precisely, we must collect four times as large a test set. To reduce our test error by a factor of one hundred, we must collect ten thousand times as large a test set. In general, such a rate of $\mathcal{O}(1/\sqrt{n})$ is often the best we can hope for in statistics.

Now that we know something about the asymptotic rate at which our test error $\epsilon_{\mathcal{D}}(f)$ converges to the true error $\epsilon(f)$, we can zoom in on some important details. Recall that the random variable of interest $\mathbf{1}(f(X) \neq Y)$ can only take values 0 and 1 and thus is a Bernoulli random variable, characterized by a parameter indicating the probability that it takes value 1. Here, 1 means that our classifier made an error, so the parameter of our random variable is actually the true error rate $\epsilon(f)$. The variance $\sigma^2$ of a Bernoulli depends on its parameter (here, $\epsilon(f)$) according to the expression $\epsilon(f)(1-\epsilon(f))$. While $\epsilon(f)$ is initially unknown, we know that it cannot be greater than 1. A little investigation of this function reveals that our variance is highest when the true error rate is close to 0.5 and can be far lower when it is close to 0 or close to 1. This tells us that the asymptotic standard deviation of our estimate $\epsilon_{\mathcal{D}}(f)$ of the error $\epsilon(f)$ (over the choice of the $n$ test samples) cannot be any greater than $\sqrt{0.25/n}$.

If we ignore the fact that this rate characterizes behavior as the test set size approaches infinity rather than when we possess finite samples, this tells us that if we want our test error $\epsilon_{\mathcal{D}}(f)$ to approximate the population error $\epsilon(f)$ such that one standard deviation corresponds to an interval of $\pm 0.01$, then we should collect roughly 2500 samples. If we want to fit two standard deviations in that range and thus be 95% confident that $\epsilon_{\mathcal{D}}(f) \in \epsilon(f) \pm 0.01$, then we will need 10,000 samples!

This turns out to be the size of the test sets for many popular benchmarks in machine learning. You might be surprised to find out that thousands of applied deep learning papers get published every year making a big deal out of error rate improvements of 0.01 or less. Of course, when the error rates are much closer to 0, then an improvement of 0.01 can indeed be a big deal.

One pesky feature of our analysis thus far is that it really only tells us about asymptotics, i.e., how the relationship between $\epsilon_{\mathcal{D}}$ and $\epsilon$ evolves as our sample size goes to infinity. Fortunately, because our random variable is bounded, we can obtain valid finite sample bounds by applying an inequality due to Hoeffding (1963):

$$P(\epsilon_{\mathcal{D}}(f) - \epsilon(f) \geq t) < \exp\left(-2nt^2\right). \tag{4.6.3}$$

Solving for the smallest dataset size that would allow us to conclude with 95% confidence that the distance $t$ between our estimate $\epsilon_{\mathcal{D}}(f)$ and the true error rate $\epsilon(f)$ does not exceed 0.01, you will find that roughly 15,000 examples are required as compared to the 10,000 examples suggested by the asymptotic analysis above. If you go deeper into statistics you will find that this trend holds generally. Guarantees that hold even in finite samples are typically slightly more conservative. Note that in the scheme of things, these numbers are not so far apart, reflecting the general usefulness of asymptotic analysis for giving us ballpark figures even if they are not guarantees we can take to court.

### 4.6.2 Test Set Reuse

In some sense, you are now set up to succeed at conducting empirical machine learning research. Nearly all practical models are developed and validated based on test set performance and you are now a master of the test set. For any fixed classifier $f$, you know how to evaluate its test error $\epsilon_{\mathcal{D}}(f)$, and know precisely what can (and cannot) be said about its population error $\epsilon(f)$.

So let's say that you take this knowledge and prepare to train your first model $f_1$. Knowing just how confident you need to be in the performance of your classifier's error rate you apply our analysis above to determine an appropriate number of examples to set aside for the test set. Moreover, let's assume that you took the lessons from Section 3.6 to heart and made sure to preserve the sanctity of the test set by conducting all of your preliminary analysis, hyperparameter tuning, and even selection among multiple competing model architectures on a validation set. Finally you evaluate your model $f_1$ on the test set and report an unbiased estimate of the population error with an associated confidence interval.

So far everything seems to be going well. However, that night you wake up at 3am with a brilliant idea for a new modeling approach. The next day, you code up your new model, tune its hyperparameters on the validation set and not only are you getting your new model $f_2$ to work but its error rate appears to be much lower than $f_1$'s. However, the thrill of discovery suddenly fades as you prepare for the final evaluation. You do not have a test set!

Even though the original test set $\mathcal{D}$ is still sitting on your server, you now face two formidable problems. First, when you collected your test set, you determined the required level of precision under the assumption that you were evaluating a single classifier $f$. However, if you get into the business of evaluating multiple classifiers $f_1, ..., f_k$ on the same test set, you must consider the problem of false discovery. Before, you might have been 95% sure that $\epsilon_{\mathcal{D}}(f) \in \epsilon(f) \pm 0.01$ for a single classifier $f$ and thus the probability of a misleading result was a mere 5%. With $k$ classifiers in the mix, it can be hard to guarantee that there is not even one among them whose test set performance is misleading. With 20 classifiers under consideration, you might have no power at all to rule out the possibility that at least one among them received a misleading score. This problem relates to multiple hypothesis testing, which despite a vast literature in statistics, remains a persistent problem plaguing scientific research.

If that is not enough to worry you, there is a special reason to distrust the results that you get on subsequent evaluations. Recall that our analysis of test set performance rested on the assumption that the classifier was chosen absent any contact with the test set and thus we could view the test set as drawn randomly from the underlying population. Here, not only are you testing multiple functions, the subsequent function $f_2$ was chosen after you observed the test set performance of $f_1$. Once information from the test set has leaked to the modeler, it can never be a true test set again in the strictest sense. This problem is called *adaptive overfitting* and has recently emerged as a topic of intense interest to learning theorists and statisticians (Dwork *et al.*, 2015). Fortunately, while it is possible to leak all information out of a holdout set, and the theoretical worst case scenarios are bleak, these analyses may be too conservative. In practice, take care to create real test sets, to consult them as infrequently as possible, to account for multiple hypothesis testing when reporting confidence intervals, and to dial up your vigilance more aggressively when the stakes are high and your dataset size is small. When running a series of benchmark challenges, it is often good practice to maintain several test sets so that after each round, the old test set can be demoted to a validation set.

### 4.6.3 Statistical Learning Theory

Put simply, *test sets are all that we really have*, and yet this fact seems strangely unsatisfying. First, we seldom possess a *true test set*—unless we are the ones creating the dataset, someone else has probably already evaluated their own classifier on our ostensible "test set". And even when we have first dibs, we soon find ourselves frustrated, wishing we could evaluate our subsequent modeling attempts without the gnawing feeling that we cannot trust our numbers. Moreover, even a true test set can only tell us *post hoc* whether a classifier has in fact generalized to the population, not whether we have any reason to expect *a priori* that it should generalize.

With these misgivings in mind, you might now be sufficiently primed to see the appeal of *statistical learning theory*, the mathematical subfield of machine learning whose practitioners aim to elucidate the fundamental principles that explain why/when models trained on empirical data can/will generalize to unseen data. One of the primary aims of statistical learning researchers has been to bound the generalization gap, relating the properties of the model class to the number of samples in the dataset.

Learning theorists aim to bound the difference between the *empirical error* $\epsilon_\mathcal{S}(f_\mathcal{S})$ of a learned classifier $f_\mathcal{S}$, both trained and evaluated on the training set $\mathcal{S}$, and the true error $\epsilon(f_\mathcal{S})$ of that same classifier on the underlying population. This might look similar to the evaluation problem that we just addressed but there is a major difference. Earlier, the classifier $f$ was fixed and we only needed a dataset for evaluative purposes. And indeed, any fixed classifier does generalize: its error on a (previously unseen) dataset is an unbiased estimate of the population error. But what can we say when a classifier is trained and evaluated on the same dataset? Can we ever be confident that the training error will be close to the testing error?

Suppose that our learned classifier $f_\mathcal{S}$ must be chosen from some pre-specified set of functions $\mathcal{F}$. Recall from our discussion of test sets that while it is easy to estimate the error of a single classifier, things get hairy when we begin to consider collections of classifiers. Even if the empirical error of any one (fixed) classifier will be close to its true error with high probability, once we consider a collection of classifiers, we need to worry about the possibility that *just one* of them will receive a badly estimated error. The worry is that we might pick such a classifier and thereby grossly underestimate the population error. Moreover, even for linear models, because their parameters are continuously valued, we are typically choosing from an infinite class of functions ($|\mathcal{F}| = \infty$).

One ambitious solution to the problem is to develop analytic tools for proving uniform convergence, i.e., that with high probability, the empirical error rate for every classifier in the class $f \in \mathcal{F}$ will *simultaneously* converge to its true error rate. In other words, we seek a theoretical principle that would allow us to state that with probability at least $1 - \delta$ (for some small $\delta$) no classifier's error rate $\epsilon(f)$ (among all classifiers in the class $\mathcal{F}$) will be misestimated by more than some small amount $\alpha$. Clearly, we cannot make such statements for all model classes $\mathcal{F}$. Recall the class of memorization machines that always achieve empirical error $0$ but never outperform random guessing on the underlying population.

In a sense the class of memorizers is too flexible. No such a uniform convergence result could possibly hold. On the other hand, a fixed classifier is useless—it generalizes perfectly, but fits neither the training data nor the test data. The central question of learning has thus historically been framed as a trade-off between more flexible (higher variance) model classes that better fit the training data but risk overfitting, versus more rigid (higher bias) model classes that generalize well but risk underfitting. A central question in learning theory has been to develop the appropriate mathematical analysis to quantify where a model sits along this spectrum, and to provide the associated guarantees.

In a series of seminal papers, Vapnik and Chervonenkis extended the theory on the convergence of relative frequencies to more general classes of functions (Vapnik and Chervonenkis, 1964, Vapnik and Chervonenkis, 1968, Vapnik and Chervonenkis, 1971, Vapnik and Chervonenkis, 1981, Vapnik and Chervonenkis, 1991, Vapnik and Chervonenkis, 1974). One of the key contributions of this line of work is the Vapnik–Chervonenkis (VC) dimension, which measures (one notion of) the complexity (flexibility) of a model class. Moreover, one of their key results bounds the difference between the empirical error and the population error as a function of the VC dimension and the number of samples:

$$P\left(R[p, f] - R_{\textrm{emp}}[\mathbf{X}, \mathbf{Y}, f] < \alpha\right) \geq 1-\delta \ \textrm{ for } \ \alpha \geq c \sqrt{(\textrm{VC} - \log \delta)/n}. \tag{4.6.4}$$

Here $\delta > 0$ is the probability that the bound is violated, $\alpha$ is the upper bound on the generalization gap, and $n$ is the dataset size. Lastly, $c > 0$ is a constant that depends only on the scale of the loss that can be incurred. One use of the bound might be to plug in desired values of $\delta$ and $\alpha$ to determine how many samples to collect. The VC dimension quantifies the largest number of data points for which we can assign any arbitrary (binary) labeling and for each find some model $f$ in the class that agrees with that labeling. For example, linear models on $d$-dimensional inputs have VC dimension $d + 1$. It is easy to see that a line can assign any possible labeling to three points in two dimensions, but not to four. Unfortunately, the theory tends to be overly pessimistic for more complex models and obtaining this guarantee typically requires far more examples than are actually needed to achieve the desired error rate. Note also that fixing the model class and $\delta$, our error rate again decays with the usual $\mathcal{O}(1/\sqrt{n})$ rate. It seems unlikely that we could do better in terms of $n$. However, as we vary the model class, VC dimension can present a pessimistic picture of the generalization gap.

### 4.6.4 Summary

The most straightforward way to evaluate a model is to consult a test set comprised of previously unseen data. Test set evaluations provide an unbiased estimate of the true error and converge at the desired $\mathcal{O}(1/\sqrt{n})$ rate as the test set grows. We can provide approximate confidence intervals based on exact asymptotic distributions or valid finite sample confidence intervals based on (more conservative) finite sample guarantees. Indeed test set evaluation is the bedrock of modern machine learning research. However, test sets are seldom true test sets (used by multiple researchers again and again). Once the same test set is used to evaluate multiple models, controlling for false discovery can be difficult. This can cause huge problems in theory. In practice, the significance of the problem depends on the size of the holdout sets in question and whether they are merely being used to choose hyperparameters or if they are leaking information more directly. Nevertheless, it is good practice to curate real test sets (or multiple) and to be as conservative as possible about how often they are used.

Hoping to provide a more satisfying solution, statistical learning theorists have developed methods for guaranteeing uniform convergence over a model class. If indeed every model's empirical error simultaneously converges to its true error, then we are free to choose the model that performs best, minimizing the training error, knowing that it too will perform similarly well on the holdout data. Crucially, any one of such results must depend on some property of the model class. Vladimir Vapnik and Alexey Chernovenkis introduced the VC dimension, presenting uniform convergence results that hold for all models in a VC class. The training errors for all models in the class are (simultaneously) guaranteed to be close to their true errors, and guaranteed to grow even closer at $O(1/\sqrt{n})$ rates. Following the revolutionary discovery of VC dimension, numerous alternative complexity measures have been proposed, each facilitating an analogous generalization guarantee. See Boucheron *et al.* (2005) for a detailed discussion of several advanced ways of measuring function complexity. Unfortunately, while these complexity measures have become broadly useful tools in statistical theory, they turn out to be powerless (as straightforwardly applied) for explaining why deep neural networks generalize. Deep neural networks often have millions of parameters (or more), and can easily assign random labels to large collections of points. Nevertheless, they generalize well on practical problems and, surprisingly, they often generalize better, when they are larger and deeper, despite incurring higher VC dimensions. In the next chapter, we will revisit generalization in the context of deep learning.

## 4.6.5 Exercises

1. If we wish to estimate the error of a fixed model $f$ to within $0.0001$ with probability greater than 99.9%, how many samples do we need?

2. Suppose that somebody else possesses a labeled test set $\mathcal{D}$ and only makes available the unlabeled inputs (features). Now suppose that you can only access the test set labels by running a model $f$ (with no restrictions placed on the model class) on each of the unlabeled inputs and receiving the corresponding error $\epsilon_{\mathcal{D}}(f)$. How many models would you need to evaluate before you leak the entire test set and thus could appear to have error 0, regardless of your true error?

3. What is the VC dimension of the class of fifth-order polynomials?

4. What is the VC dimension of axis-aligned rectangles on two-dimensional data?

Discussions[100].

# 4.7 Environment and Distribution Shift

In the previous sections, we worked through a number of hands-on applications of machine learning, fitting models to a variety of datasets. And yet, we never stopped to contemplate either where data came from in the first place or what we ultimately plan to do with the outputs from our models. Too often, machine learning developers in possession of data rush to develop models without pausing to consider these fundamental issues.

Many failed machine learning deployments can be traced back to this failure. Sometimes models appear to perform marvelously as measured by test set accuracy but fail catastrophically in deployment when the distribution of data suddenly shifts. More insidiously, sometimes the very deployment of a model can be the catalyst that perturbs the data distribution. Say, for example, that we trained a model to predict who will repay rather than default on a loan, finding that an applicant's choice of footwear was associated with the risk of default (Oxfords indicate repayment, sneakers indicate default). We might be inclined thereafter to grant a loan to any applicant wearing Oxfords and to deny all applicants wearing sneakers.

In this case, our ill-considered leap from pattern recognition to decision-making and our failure to critically consider the environment might have disastrous consequences. For starters, as soon as we began making decisions based on footwear, customers would catch on and change their behavior. Before long, all applicants would be wearing Oxfords, without any coincident improvement in credit-worthiness. Take a minute to digest this because similar issues abound in many applications of machine learning: by introducing our model-based decisions to the environment, we might break the model.

While we cannot possibly give these topics a complete treatment in one section, we aim here to expose some common concerns, and to stimulate the critical thinking required to detect such situations early, mitigate damage, and use machine learning responsibly. Some of the solutions are simple (ask for the "right" data), some are technically difficult (implement a reinforcement learning system), and others require that we step outside the realm of statistical prediction altogether and grapple with difficult philosophical questions concerning the ethical application of algorithms.

## 4.7.1 Types of Distribution Shift

To begin, we stick with the passive prediction setting considering the various ways that data distributions might shift and what might be done to salvage model performance. In one classic setup, we assume that our training data was sampled from some distribution $p_S(\mathbf{x}, y)$ but that our test data will consist of unlabeled examples drawn from some different distribution $p_T(\mathbf{x}, y)$. Already, we must confront a sobering reality. Absent any assumptions on how $p_S$ and $p_T$ relate to each other, learning a robust classifier is impossible.

Consider a binary classification problem, where we wish to distinguish between dogs and cats. If the distribution can shift in arbitrary ways, then our setup permits the pathological case in which the distribution over inputs remains constant: $p_S(\mathbf{x}) = p_T(\mathbf{x})$, but the labels are all flipped: $p_S(y \mid \mathbf{x}) = 1 - p_T(y \mid \mathbf{x})$. In other words, if God can suddenly decide that in the future all "cats" are now dogs and what we previously called "dogs" are now cats—without any change in the distribution of inputs $p(\mathbf{x})$, then we cannot possibly distinguish this setting from one in which the distribution did not change at all.

Fortunately, under some restricted assumptions on the ways our data might change in the future, principled algorithms can detect shift and sometimes even adapt on the fly, improving on the accuracy of the original classifier.

## Covariate Shift

Among categories of distribution shift, covariate shift may be the most widely studied. Here, we assume that while the distribution of inputs may change over time, the labeling function, i.e., the conditional distribution $P(y \mid \mathbf{x})$ does not change. Statisticians call this *covariate shift* because the problem arises due to a shift in the distribution of the covariates (features). While we can sometimes reason about distribution shift without invoking causality, we note that covariate shift is the natural assumption to invoke in settings where we believe that $\mathbf{x}$ causes $y$.

Consider the challenge of distinguishing cats and dogs. Our training data might consist of images of the kind in Fig. 4.7.1.

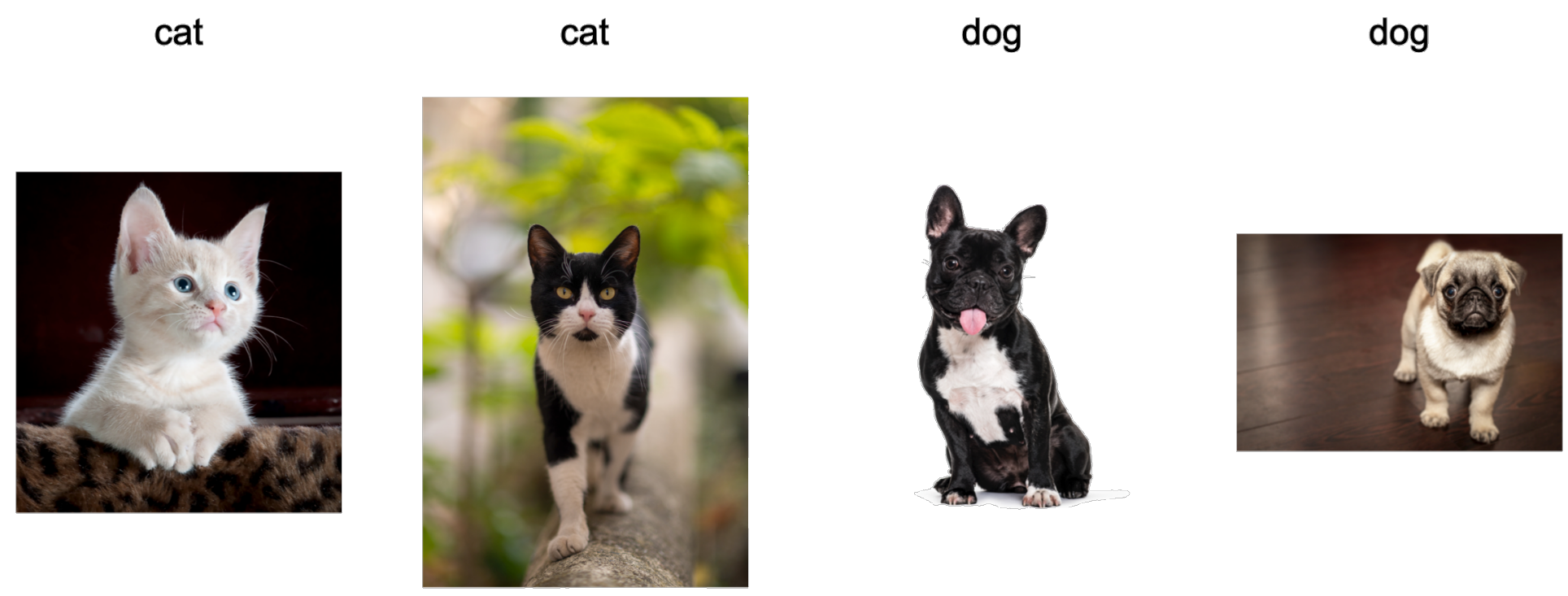

Fig. 4.7.1 Training data for distinguishing cats and dogs (illustrations: Lafeez Hossain / 500px / Getty Images; ilkermetinkursova / iStock / Getty Images Plus; GlobalP / iStock / Getty Images Plus; Musthafa Aboobakuru / 500px / Getty Images).

At test time we are asked to classify the images in Fig. 4.7.2.

The training set consists of photos, while the test set contains only cartoons. Training on a dataset with substantially different characteristics from the test set can spell trouble absent a coherent plan for how to adapt to the new domain.

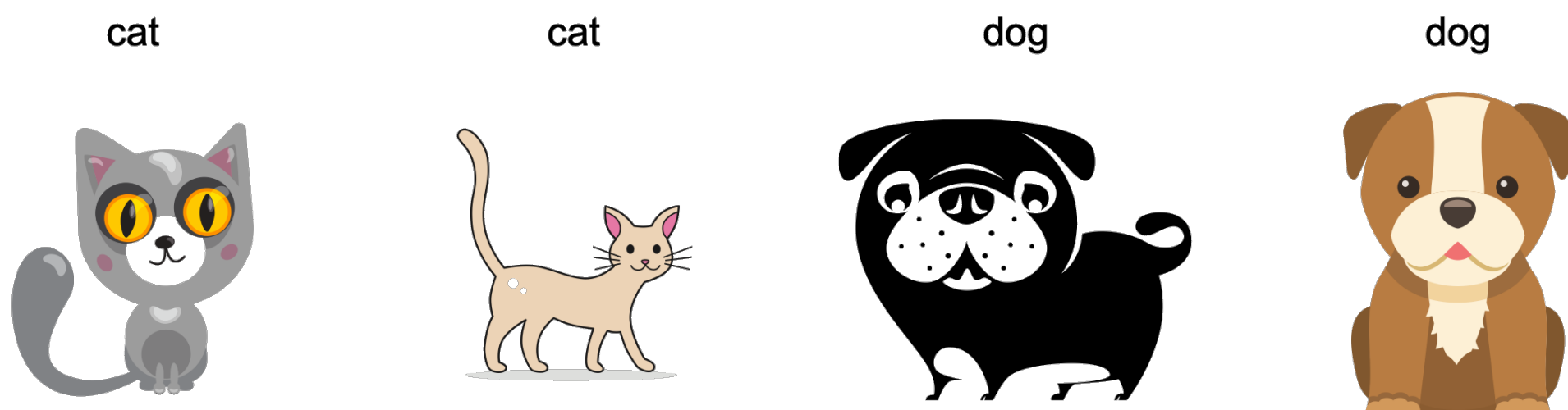

Fig. 4.7.2 Test data for distinguishing cats and dogs (illustrations: SIBAS_minich / iStock / Getty Images Plus; Ghrzuzudu / iStock / Getty Images Plus; id-work / DigitalVision Vectors / Getty Images; Yime / iStock / Getty Images Plus).

### Label Shift

*Label shift* describes the converse problem. Here, we assume that the label marginal $P(y)$ can change but the class-conditional distribution $P(\mathbf{x} \mid y)$ remains fixed across domains. Label shift is a reasonable assumption to make when we believe that $y$ causes $\mathbf{x}$. For example, we may want to predict diagnoses given their symptoms (or other manifestations), even as the relative prevalence of diagnoses are changing over time. Label shift is the appropriate assumption here because diseases cause symptoms. In some degenerate cases the label shift and covariate shift assumptions can hold simultaneously. For example, when the label is deterministic, the covariate shift assumption will be satisfied, even when $y$ causes $\mathbf{x}$. Interestingly, in these cases, it is often advantageous to work with methods that flow from the label shift assumption. That is because these methods tend to involve manipulating objects that look like labels (often low-dimensional), as opposed to objects that look like inputs, which tend to be high-dimensional in deep learning.

### Concept Shift

We may also encounter the related problem of *concept shift*, which arises when the very definitions of labels can change. This sounds weird—a *cat* is a *cat*, no? However, other categories are subject to changes in usage over time. Diagnostic criteria for mental illness, what passes for fashionable, and job titles, are all subject to considerable amounts of concept shift. It turns out that if we navigate around the United States, shifting the source of our data by geography, we will find considerable concept shift regarding the distribution of names for *soft drinks* as shown in Fig. 4.7.3.

If we were to build a machine translation system, the distribution $P(y \mid \mathbf{x})$ might be different depending on our location. This problem can be tricky to spot. We might hope to exploit knowledge that shift only takes place gradually either in a temporal or geographic sense.

## 4.7.2 Examples of Distribution Shift

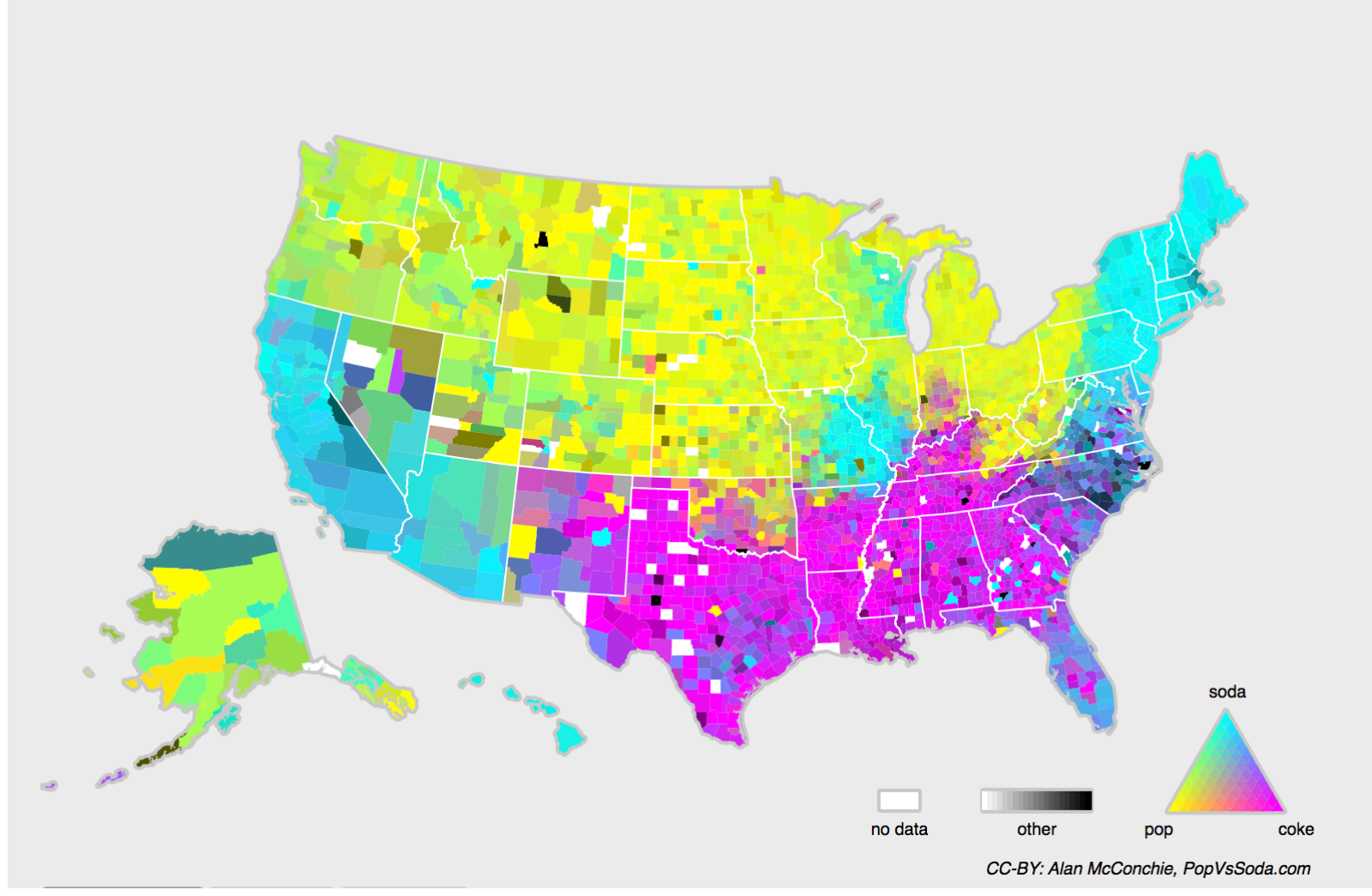

Fig. 4.7.3 Concept shift for soft drink names in the United States (CC-BY: Alan McConchie, PopVsSoda.com).

---

Before delving into formalism and algorithms, we can discuss some concrete situations where covariate or concept shift might not be obvious.

### Medical Diagnostics

Imagine that you want to design an algorithm to detect cancer. You collect data from healthy and sick people and you train your algorithm. It works fine, giving you high accuracy and you conclude that you are ready for a successful career in medical diagnostics. *Not so fast.*

The distributions that gave rise to the training data and those you will encounter in the wild might differ considerably. This happened to an unfortunate startup that some of we authors worked with years ago. They were developing a blood test for a disease that predominantly affects older men and hoped to study it using blood samples that they had collected from patients. However, it is considerably more difficult to obtain blood samples from healthy men than from sick patients already in the system. To compensate, the startup solicited blood donations from students on a university campus to serve as healthy controls in developing their test. Then they asked whether we could help them to build a classifier for detecting the disease.

As we explained to them, it would indeed be easy to distinguish between the healthy and sick cohorts with near-perfect accuracy. However, that is because the test subjects differed in age, hormone levels, physical activity, diet, alcohol consumption, and many more factors unrelated to the disease. This was unlikely to be the case with real patients. Due to their sampling procedure, we could expect to encounter extreme covariate shift. Moreover, this case was unlikely to be correctable via conventional methods. In short, they wasted a significant sum of money.

### Self-Driving Cars

Say a company wanted to leverage machine learning for developing self-driving cars. One key component here is a roadside detector. Since real annotated data is expensive to get, they had the (smart and questionable) idea to use synthetic data from a game rendering engine as additional training data. This worked really well on "test data" drawn from the rendering engine. Alas, inside a real car it was a disaster. As it turned out, the roadside had been rendered with a very simplistic texture. More importantly, *all* the roadside had been rendered with the *same* texture and the roadside detector learned about this "feature" very quickly.

A similar thing happened to the US Army when they first tried to detect tanks in the forest. They took aerial photographs of the forest without tanks, then drove the tanks into the forest and took another set of pictures. The classifier appeared to work *perfectly*. Unfortunately, it had merely learned how to distinguish trees with shadows from trees without shadows—the first set of pictures was taken in the early morning, the second set at noon.

### Nonstationary Distributions

A much more subtle situation arises when the distribution changes slowly (also known as *nonstationary distribution*) and the model is not updated adequately. Below are some typical cases.

- We train a computational advertising model and then fail to update it frequently (e.g., we forget to incorporate that an obscure new device called an iPad was just launched).
- We build a spam filter. It works well at detecting all spam that we have seen so far. But then the spammers wise up and craft new messages that look unlike anything we have seen before.
- We build a product recommendation system. It works throughout the winter but then continues to recommend Santa hats long after Christmas.

### More Anecdotes

- We build a face detector. It works well on all benchmarks. Unfortunately it fails on test data—the offending examples are close-ups where the face fills the entire image (no such data was in the training set).
- We build a web search engine for the US market and want to deploy it in the UK.
- We train an image classifier by compiling a large dataset where each among a large set of classes is equally represented in the dataset, say 1000 categories, represented by 1000 images each. Then we deploy the system in the real world, where the actual label distribution of photographs is decidedly non-uniform.

## 4.7.3 Correction of Distribution Shift

As we have discussed, there are many cases where training and test distributions $P(\mathbf{x}, y)$ are different. In some cases, we get lucky and the models work despite covariate, label, or concept shift. In other cases, we can do better by employing principled strategies to cope with the shift. The remainder of this section grows considerably more technical. The impatient reader could continue on to the next section as this material is not prerequisite to subsequent concepts.

### Empirical Risk and Risk

Let's first reflect on what exactly is happening during model training: we iterate over features and associated labels of training data $\{(\mathbf{x}_1, y_1), \ldots, (\mathbf{x}_n, y_n)\}$ and update the parameters of a model $f$ after every minibatch. For simplicity we do not consider regularization, so we largely minimize the loss on the training:

$$\underset{f}{\text{minimize}} \frac{1}{n} \sum_{i=1}^{n} l(f(\mathbf{x}_i), y_i), \tag{4.7.1}$$

where $l$ is the loss function measuring "how bad" the prediction $f(\mathbf{x}_i)$ is given the associated label $y_i$. Statisticians call the term in (4.7.1) *empirical risk*. The *empirical risk* is an average loss over the training data for approximating the *risk*, which is the expectation of the loss over the entire population of data drawn from their true distribution $p(\mathbf{x}, y)$:

$$E_{p(\mathbf{x},y)}[l(f(\mathbf{x}), y)] = \int\int l(f(\mathbf{x}), y) p(\mathbf{x}, y) \; d\mathbf{x}dy. \tag{4.7.2}$$

However, in practice we typically cannot obtain the entire population of data. Thus, *empirical risk minimization*, which is minimizing the empirical risk in (4.7.1), is a practical strategy for machine learning, with the hope of approximately minimizing the risk.

### Covariate Shift Correction

Assume that we want to estimate some dependency $P(y \mid \mathbf{x})$ for which we have labeled data $(\mathbf{x}_i, y_i)$. Unfortunately, the observations $\mathbf{x}_i$ are drawn from some *source distribution* $q(\mathbf{x})$ rather than the *target distribution* $p(\mathbf{x})$. Fortunately, the dependency assumption means that the conditional distribution does not change: $p(y \mid \mathbf{x}) = q(y \mid \mathbf{x})$. If the source distribution $q(\mathbf{x})$ is "wrong", we can correct for that by using the following simple identity in the risk:

$$\int\int l(f(\mathbf{x}), y) p(y \mid \mathbf{x}) p(\mathbf{x}) \; d\mathbf{x}dy = \int\int l(f(\mathbf{x}), y) q(y \mid \mathbf{x}) q(\mathbf{x}) \frac{p(\mathbf{x})}{q(\mathbf{x})} \; d\mathbf{x}dy. \tag{4.7.3}$$

In other words, we need to reweigh each data example by the ratio of the probability that it would have been drawn from the correct distribution to that from the wrong one:

$$\beta_i \stackrel{\text{def}}{=} \frac{p(\mathbf{x}_i)}{q(\mathbf{x}_i)}. \tag{4.7.4}$$

Plugging in the weight $\beta_i$ for each data example $(\mathbf{x}_i, y_i)$ we can train our model using *weighted empirical risk minimization*:

$$\underset{f}{\operatorname{minimize}} \frac{1}{n} \sum_{i=1}^{n} \beta_i l(f(\mathbf{x}_i), y_i). \tag{4.7.5}$$

Alas, we do not know that ratio, so before we can do anything useful we need to estimate it. Many methods are available, including some fancy operator-theoretic approaches that attempt to recalibrate the expectation operator directly using a minimum-norm or a maximum entropy principle. Note that for any such approach, we need samples drawn from both distributions—the "true" $p$, e.g., by access to test data, and the one used for generating the training set $q$ (the latter is trivially available). Note however, that we only need features $\mathbf{x} \sim p(\mathbf{x})$; we do not need to access labels $y \sim p(y)$.

In this case, there exists a very effective approach that will give almost as good results as the original: namely, logistic regression, which is a special case of softmax regression (see Section 4.1) for binary classification. This is all that is needed to compute estimated probability ratios. We learn a classifier to distinguish between data drawn from $p(\mathbf{x})$ and data drawn from $q(\mathbf{x})$. If it is impossible to distinguish between the two distributions then it means that the associated instances are equally likely to come from either one of those two distributions. On the other hand, any instances that can be well discriminated should be significantly overweighted or underweighted accordingly.

For simplicity's sake assume that we have an equal number of instances from both distributions $p(\mathbf{x})$ and $q(\mathbf{x})$, respectively. Now denote by $z$ labels that are 1 for data drawn from $p$ and $-1$ for data drawn from $q$. Then the probability in a mixed dataset is given by

$$P(z = 1 \mid \mathbf{x}) = \frac{p(\mathbf{x})}{p(\mathbf{x}) + q(\mathbf{x})} \text{ and hence } \frac{P(z = 1 \mid \mathbf{x})}{P(z = -1 \mid \mathbf{x})} = \frac{p(\mathbf{x})}{q(\mathbf{x})}. \tag{4.7.6}$$

Thus, if we use a logistic regression approach, where $P(z = 1 \mid \mathbf{x}) = \frac{1}{1+\exp(-h(\mathbf{x}))}$ ($h$ is a parametrized function), it follows that

$$\beta_i = \frac{1/(1 + \exp(-h(\mathbf{x}_i)))}{\exp(-h(\mathbf{x}_i))/(1 + \exp(-h(\mathbf{x}_i)))} = \exp(h(\mathbf{x}_i)). \tag{4.7.7}$$

As a result, we need to solve two problems: the first, to distinguish between data drawn from both distributions, and then a weighted empirical risk minimization problem in (4.7.5) where we weigh terms by $\beta_i$.

Now we are ready to describe a correction algorithm. Suppose that we have a training set $\{(\mathbf{x}_1, y_1), \ldots, (\mathbf{x}_n, y_n)\}$ and an unlabeled test set $\{\mathbf{u}_1, \ldots, \mathbf{u}_m\}$. For covariate shift, we assume that $\mathbf{x}_i$ for all $1 \leq i \leq n$ are drawn from some source distribution and $\mathbf{u}_i$ for all $1 \leq i \leq m$ are drawn from the target distribution. Here is a prototypical algorithm for correcting covariate shift:

1. Create a binary-classification training set: $\{(\mathbf{x}_1, -1), \ldots, (\mathbf{x}_n, -1), (\mathbf{u}_1, 1), \ldots, (\mathbf{u}_m, 1)\}$.
2. Train a binary classifier using logistic regression to get the function $h$.
3. Weigh training data using $\beta_i = \exp(h(\mathbf{x}_i))$ or better $\beta_i = \min(\exp(h(\mathbf{x}_i)), c)$ for some constant $c$.
4. Use weights $\beta_i$ for training on $\{(\mathbf{x}_1, y_1), \ldots, (\mathbf{x}_n, y_n)\}$ in (4.7.5).

Note that the above algorithm relies on a crucial assumption. For this scheme to work, we need that each data example in the target (e.g., test time) distribution had nonzero probability of occurring at training time. If we find a point where $p(\mathbf{x}) > 0$ but $q(\mathbf{x}) = 0$, then the corresponding importance weight should be infinity.

## Label Shift Correction

Assume that we are dealing with a classification task with $k$ categories. Using the same notation in Section 4.7.3, $q$ and $p$ are the source distribution (e.g., training time) and target distribution (e.g., test time), respectively. Assume that the distribution of labels shifts over time: $q(y) \neq p(y)$, but the class-conditional distribution stays the same: $q(\mathbf{x} \mid y) = p(\mathbf{x} \mid y)$. If the source distribution $q(y)$ is "wrong", we can correct for that according to the following identity in the risk as defined in (4.7.2):

$$\int\int l(f(\mathbf{x}), y) p(\mathbf{x} \mid y) p(y) \; d\mathbf{x}dy = \int\int l(f(\mathbf{x}), y) q(\mathbf{x} \mid y) q(y) \frac{p(y)}{q(y)} \; d\mathbf{x}dy. \tag{4.7.8}$$

Here, our importance weights will correspond to the label likelihood ratios:

$$\beta_i \stackrel{\text{def}}{=} \frac{p(y_i)}{q(y_i)}. \tag{4.7.9}$$

One nice thing about label shift is that if we have a reasonably good model on the source distribution, then we can get consistent estimates of these weights without ever having to deal with the ambient dimension. In deep learning, the inputs tend to be high-dimensional objects like images, while the labels are often simpler objects like categories.

To estimate the target label distribution, we first take our reasonably good off-the-shelf classifier (typically trained on the training data) and compute its "confusion" matrix using the validation set (also from the training distribution). The *confusion matrix*, $\mathbf{C}$, is simply a $k \times k$ matrix, where each column corresponds to the label category (ground truth) and each row corresponds to our model's predicted category. Each cell's value $c_{ij}$ is the fraction of total predictions on the validation set where the true label was $j$ and our model predicted $i$.

Now, we cannot calculate the confusion matrix on the target data directly because we do not get to see the labels for the examples that we see in the wild, unless we invest in a complex real-time annotation pipeline. What we can do, however, is average all of our model's predictions at test time together, yielding the mean model outputs $\mu(\hat{\mathbf{y}}) \in \mathbb{R}^k$, where the $i^{\text{th}}$ element $\mu(\hat{y}_i)$ is the fraction of the total predictions on the test set where our model predicted $i$.

It turns out that under some mild conditions—if our classifier was reasonably accurate in the first place, and if the target data contains only categories that we have seen before, and if the label shift assumption holds in the first place (the strongest assumption here)—we can estimate the test set label distribution by solving a simple linear system

$$\mathbf{C}p(\mathbf{y}) = \mu(\hat{\mathbf{y}}), \tag{4.7.10}$$

because as an estimate $\sum_{j=1}^{k} c_{ij} p(y_j) = \mu(\hat{y}_i)$ holds for all $1 \leq i \leq k$, where $p(y_j)$ is the $j^\mathrm{th}$ element of the $k$-dimensional label distribution vector $p(\mathbf{y})$. If our classifier is sufficiently accurate to begin with, then the confusion matrix $\mathbf{C}$ will be invertible, and we get a solution $p(\mathbf{y}) = \mathbf{C}^{-1}\mu(\hat{\mathbf{y}})$.

Because we observe the labels on the source data, it is easy to estimate the distribution $q(y)$. Then, for any training example $i$ with label $y_i$, we can take the ratio of our estimated $p(y_i)/q(y_i)$ to calculate the weight $\beta_i$, and plug this into weighted empirical risk minimization in (4.7.5).

### Concept Shift Correction

Concept shift is much harder to fix in a principled manner. For instance, in a situation where suddenly the problem changes from distinguishing cats from dogs to one of distinguishing white from black animals, it will be unreasonable to assume that we can do much better than just collecting new labels and training from scratch. Fortunately, in practice, such extreme shifts are rare. Instead, what usually happens is that the task keeps on changing slowly. To make things more concrete, here are some examples:

- In computational advertising, new products are launched, old products become less popular. This means that the distribution over ads and their popularity changes gradually and any click-through rate predictor needs to change gradually with it.
- Traffic camera lenses degrade gradually due to environmental wear, affecting image quality progressively.
- News content changes gradually (i.e., most of the news remains unchanged but new stories appear).

In such cases, we can use the same approach that we used for training networks to make them adapt to the change in the data. In other words, we use the existing network weights and simply perform a few update steps with the new data rather than training from scratch.

## 4.7.4 A Taxonomy of Learning Problems

Armed with knowledge about how to deal with changes in distributions, we can now consider some other aspects of machine learning problem formulation.

### Batch Learning

In *batch learning*, we have access to training features and labels $\{(\mathbf{x}_1, y_1), \ldots, (\mathbf{x}_n, y_n)\}$, which we use to train a model $f(\mathbf{x})$. Later on, we deploy this model to score new data $(\mathbf{x}, y)$ drawn from the same distribution. This is the default assumption for any of the problems that we discuss here. For instance, we might train a cat detector based on lots of pictures of cats and dogs. Once we have trained it, we ship it as part of a smart catdoor computer vision system that lets only cats in. This is then installed in a customer's home and is never updated again (barring extreme circumstances).

### Online Learning

Now imagine that the data $(\mathbf{x}_i, y_i)$ arrives one sample at a time. More specifically, assume that we first observe $\mathbf{x}_i$, then we need to come up with an estimate $f(\mathbf{x}_i)$. Only once we have done this do we observe $y_i$ and so receive a reward or incur a loss, given our decision. Many real problems fall into this category. For example, we need to predict tomorrow's stock price, which allows us to trade based on that estimate and at the end of the day we find out whether our estimate made us a profit. In other words, in *online learning*, we have the following cycle where we are continuously improving our model given new observations:

$$\begin{aligned}&\textrm{model } f_t \longrightarrow \textrm{data } \mathbf{x}_t \longrightarrow \textrm{estimate } f_t(\mathbf{x}_t) \longrightarrow\\ \textrm{obs}&\textrm{ervation } y_t \longrightarrow \textrm{loss } l(y_t, f_t(\mathbf{x}_t)) \longrightarrow \textrm{model } f_{t+1}\end{aligned} \tag{4.7.11}$$

### Bandits

*Bandits* are a special case of the problem above. While in most learning problems we have a continuously parametrized function $f$ where we want to learn its parameters (e.g., a deep network), in a *bandit* problem we only have a finite number of arms that we can pull, i.e., a finite number of actions that we can take. It is not very surprising that for this simpler problem stronger theoretical guarantees in terms of optimality can be obtained. We list it mainly since this problem is often (confusingly) treated as if it were a distinct learning setting.

### Control

In many cases the environment remembers what we did. Not necessarily in an adversarial manner but it will just remember and the response will depend on what happened before. For instance, a coffee boiler controller will observe different temperatures depending on whether it was heating the boiler previously. PID (proportional-integral-derivative) controller algorithms are a popular choice there. Likewise, a user's behavior on a news site will depend on what we showed them previously (e.g., they will read most news only once). Many such algorithms form a model of the environment in which they act so as to make their decisions appear less random. Recently, control theory (e.g., PID variants) has also been used to automatically tune hyperparameters to achieve better disentangling and reconstruction quality, and improve the diversity of generated text and the reconstruction quality of generated images (Shao *et al.*, 2020).

#### Reinforcement Learning

In the more general case of an environment with memory, we may encounter situations where the environment is trying to cooperate with us (cooperative games, in particular for non-zero-sum games), or others where the environment will try to win. Chess, Go, Backgammon, or StarCraft are some of the cases in *reinforcement learning*. Likewise, we might want to build a good controller for autonomous cars. Other cars are likely to respond to the autonomous car's driving style in nontrivial ways, e.g., trying to avoid it, trying to cause an accident, or trying to cooperate with it.

#### Considering the Environment

One key distinction between the different situations above is that a strategy that might have worked throughout in the case of a stationary environment, might not work throughout in an environment that can adapt. For instance, an arbitrage opportunity discovered by a trader is likely to disappear once it is exploited. The speed and manner at which the environment changes determines to a large extent the type of algorithms that we can bring to bear. For instance, if we know that things may only change slowly, we can force any estimate to change only slowly, too. If we know that the environment might change instantaneously, but only very infrequently, we can make allowances for that. These types of knowledge are crucial for the aspiring data scientist in dealing with concept shift, i.e., when the problem that is being solved can change over time.

### 4.7.5 Fairness, Accountability, and Transparency in Machine Learning

Finally, it is important to remember that when you deploy machine learning systems you are not merely optimizing a predictive model—you are typically providing a tool that will be used to (partially or fully) automate decisions. These technical systems can impact the lives of individuals who are subject to the resulting decisions. The leap from considering predictions to making decisions raises not only new technical questions, but also a slew of ethical questions that must be carefully considered. If we are deploying a medical diagnostic system, we need to know for which populations it may work and for which it may not. Overlooking foreseeable risks to the welfare of a subpopulation could cause us to administer inferior care. Moreover, once we contemplate decision-making systems, we must step back and reconsider how we evaluate our technology. Among other consequences of this change of scope, we will find that *accuracy* is seldom the right measure. For instance, when translating predictions into actions, we will often want to take into account the potential cost sensitivity of erring in various ways.

If one way of misclassifying an image could be perceived as a racial sleight of hand, while misclassification to a different category would be harmless, then we might want to adjust our thresholds accordingly, accounting for societal values in designing the decision-making protocol. We also want to be careful about how prediction systems can lead to feedback loops. For example, consider predictive policing systems, which allocate patrol officers to areas with high forecasted crime. It is easy to see how a worrying pattern can emerge:

1. Neighborhoods with more crime get more patrols.
2. Consequently, more crimes are discovered in these neighborhoods, entering the training data available for future iterations.
3. Exposed to more positives, the model predicts yet more crime in these neighborhoods.
4. In the next iteration, the updated model targets the same neighborhood even more heavily leading to yet more crimes discovered, etc.

Often, the various mechanisms by which a model's predictions become coupled to its training data are unaccounted for in the modeling process. This can lead to what researchers call *runaway feedback loops*. Additionally, we want to be careful about whether we are addressing the right problem in the first place. Predictive algorithms now play an outsize role in mediating the dissemination of information. Should the news that an individual encounters be determined by the set of Facebook pages they have *Liked*? These are just a few among the many pressing ethical dilemmas that you might encounter in a career in machine learning.

### 4.7.6 Summary

In many cases training and test sets do not come from the same distribution. This is called distribution shift. The risk is the expectation of the loss over the entire population of data drawn from their true distribution. However, this entire population is usually unavailable. Empirical risk is an average loss over the training data to approximate the risk. In practice, we perform empirical risk minimization.

Under the corresponding assumptions, covariate and label shift can be detected and corrected for at test time. Failure to account for this bias can become problematic at test time. In some cases, the environment may remember automated actions and respond in surprising ways. We must account for this possibility when building models and continue to monitor live systems, open to the possibility that our models and the environment will become entangled in unanticipated ways.

### 4.7.7 Exercises

1. What could happen when we change the behavior of a search engine? What might the users do? What about the advertisers?
2. Implement a covariate shift detector. Hint: build a classifier.

3. Implement a covariate shift corrector.
4. Besides distribution shift, what else could affect how the empirical risk approximates the risk?

Discussions[101].

# 5 Multilayer Perceptrons

In this chapter, we will introduce your first truly *deep* network. The simplest deep networks are called *multilayer perceptrons*, and they consist of multiple layers of neurons each fully connected to those in the layer below (from which they receive input) and those above (which they, in turn, influence). Although automatic differentiation significantly simplifies the implementation of deep learning algorithms, we will dive deep into how these gradients are calculated in deep networks. Then we will be ready to discuss issues relating to numerical stability and parameter initialization that are key to successfully training deep networks. When we train such high-capacity models we run the risk of overfitting. Thus, we will revisit regularization and generalization for deep networks. Throughout, we aim to give you a firm grasp not just of the concepts but also of the practice of using deep networks. At the end of this chapter, we apply what we have introduced so far to a real case: house price prediction. We punt matters relating to the computational performance, scalability, and efficiency of our models to subsequent chapters.

## 5.1 Multilayer Perceptrons

In Section 4.1, we introduced softmax regression, implementing the algorithm from scratch (Section 4.4) and using high-level APIs (Section 4.5). This allowed us to train classifiers capable of recognizing 10 categories of clothing from low-resolution images. Along the way, we learned how to wrangle data, coerce our outputs into a valid probability distribution, apply an appropriate loss function, and minimize it with respect to our model's parameters. Now that we have mastered these mechanics in the context of simple linear models, we can launch our exploration of deep neural networks, the comparatively rich class of models with which this book is primarily concerned.

```
%matplotlib inline
import torch
from d2l import torch as d2l
```

### 5.1.1 Hidden Layers

We described affine transformations in Section 3.1.1 as linear transformations with added bias. To begin, recall the model architecture corresponding to our softmax regression example, illustrated in Fig. 4.1.1. This model maps inputs directly to outputs via a single affine transformation, followed by a softmax operation. If our labels truly were related to the input data by a simple affine transformation, then this approach would be sufficient. However, linearity (in affine transformations) is a *strong* assumption.

### Limitations of Linear Models

For example, linearity implies the *weaker* assumption of *monotonicity*, i.e., that any increase in our feature must either always cause an increase in our model's output (if the corresponding weight is positive), or always cause a decrease in our model's output (if the corresponding weight is negative). Sometimes that makes sense. For example, if we were trying to predict whether an individual will repay a loan, we might reasonably assume that all other things being equal, an applicant with a higher income would always be more likely to repay than one with a lower income. While monotonic, this relationship likely is not linearly associated with the probability of repayment. An increase in income from $0 to $50,000 likely corresponds to a bigger increase in likelihood of repayment than an increase from $1 million to $1.05 million. One way to handle this might be to postprocess our outcome such that linearity becomes more plausible, by using the logistic map (and thus the logarithm of the probability of outcome).

Note that we can easily come up with examples that violate monotonicity. Say for example that we want to predict health as a function of body temperature. For individuals with a normal body temperature above 37°C (98.6°F), higher temperatures indicate greater risk. However, if the body temperatures drops below 37°C, lower temperatures indicate greater risk! Again, we might resolve the problem with some clever preprocessing, such as using the distance from 37°C as a feature.

But what about classifying images of cats and dogs? Should increasing the intensity of the pixel at location (13, 17) always increase (or always decrease) the likelihood that the image depicts a dog? Reliance on a linear model corresponds to the implicit assumption that the only requirement for differentiating cats and dogs is to assess the brightness of individual pixels. This approach is doomed to fail in a world where inverting an image preserves the category.

And yet despite the apparent absurdity of linearity here, as compared with our previous examples, it is less obvious that we could address the problem with a simple preprocessing fix. That is, because the significance of any pixel depends in complex ways on its context (the values of the surrounding pixels). While there might exist a representation of our data that would take into account the relevant interactions among our features, on top of which a linear model would be suitable, we simply do not know how to calculate it by hand. With deep neural networks, we used observational data to jointly learn both a representation via hidden layers and a linear predictor that acts upon that representation.

This problem of nonlinearity has been studied for at least a century (Fisher, 1925). For instance, decision trees in their most basic form use a sequence of binary decisions to decide upon class membership (Quinlan, 1993). Likewise, kernel methods have been used for many decades to model nonlinear dependencies (Aronszajn, 1950). This has found its way into nonparametric spline models (Wahba, 1990) and kernel methods (Schölkopf and Smola, 2002). It is also something that the brain solves quite naturally. After all, neurons feed into other neurons which, in turn, feed into other neurons again (Ramón y Cajal and Azoulay, 1894). Consequently we have a sequence of relatively simple transformations.

### Incorporating Hidden Layers

We can overcome the limitations of linear models by incorporating one or more hidden layers. The easiest way to do this is to stack many fully connected layers on top of one another. Each layer feeds into the layer above it, until we generate outputs. We can think of the first $L - 1$ layers as our representation and the final layer as our linear predictor. This architecture is commonly called a *multilayer perceptron*, often abbreviated as *MLP* (Fig. 5.1.1).

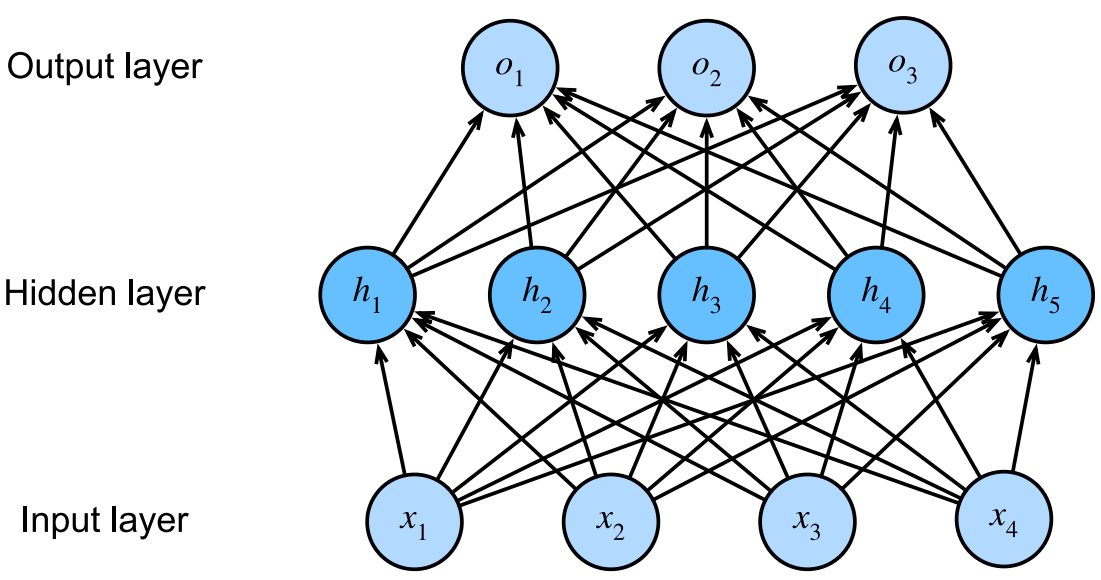

Fig. 5.1.1 An MLP with a hidden layer of five hidden units.

This MLP has four inputs, three outputs, and its hidden layer contains five hidden units. Since the input layer does not involve any calculations, producing outputs with this network requires implementing the computations for both the hidden and output layers; thus, the number of layers in this MLP is two. Note that both layers are fully connected. Every input influences every neuron in the hidden layer, and each of these in turn influences every neuron in the output layer. Alas, we are not quite done yet.

### From Linear to Nonlinear

As before, we denote by the matrix $\mathbf{X} \in \mathbb{R}^{n \times d}$ a minibatch of $n$ examples where each example has $d$ inputs (features). For a one-hidden-layer MLP whose hidden layer has $h$ hidden units, we denote by $\mathbf{H} \in \mathbb{R}^{n \times h}$ the outputs of the hidden layer, which are *hidden representations*. Since the hidden and output layers are both fully connected, we have hidden-layer weights $\mathbf{W}^{(1)} \in \mathbb{R}^{d \times h}$ and biases $\mathbf{b}^{(1)} \in \mathbb{R}^{1 \times h}$ and output-layer weights $\mathbf{W}^{(2)} \in \mathbb{R}^{h \times q}$ and biases $\mathbf{b}^{(2)} \in \mathbb{R}^{1 \times q}$. This allows us to calculate the outputs $\mathbf{O} \in \mathbb{R}^{n \times q}$ of the one-hidden-layer MLP

as follows:

$$\mathbf{H} = \mathbf{X}\mathbf{W}^{(1)} + \mathbf{b}^{(1)},$$
$$\mathbf{O} = \mathbf{H}\mathbf{W}^{(2)} + \mathbf{b}^{(2)}. \tag{5.1.1}$$

Note that after adding the hidden layer, our model now requires us to track and update additional sets of parameters. So what have we gained in exchange? You might be surprised to find out that—in the model defined above—*we gain nothing for our troubles*! The reason is plain. The hidden units above are given by an affine function of the inputs, and the outputs (pre-softmax) are just an affine function of the hidden units. An affine function of an affine function is itself an affine function. Moreover, our linear model was already capable of representing any affine function.

To see this formally we can just collapse out the hidden layer in the above definition, yielding an equivalent single-layer model with parameters $\mathbf{W} = \mathbf{W}^{(1)}\mathbf{W}^{(2)}$ and $\mathbf{b} = \mathbf{b}^{(1)}\mathbf{W}^{(2)} + \mathbf{b}^{(2)}$:

$$\mathbf{O} = (\mathbf{X}\mathbf{W}^{(1)} + \mathbf{b}^{(1)})\mathbf{W}^{(2)} + \mathbf{b}^{(2)} = \mathbf{X}\mathbf{W}^{(1)}\mathbf{W}^{(2)} + \mathbf{b}^{(1)}\mathbf{W}^{(2)} + \mathbf{b}^{(2)} = \mathbf{X}\mathbf{W} + \mathbf{b}. \tag{5.1.2}$$

In order to realize the potential of multilayer architectures, we need one more key ingredient: a nonlinear *activation function* $\sigma$ to be applied to each hidden unit following the affine transformation. For instance, a popular choice is the ReLU (rectified linear unit) activation function (Nair and Hinton, 2010) $\sigma(x) = \max(0, x)$ operating on its arguments elementwise. The outputs of activation functions $\sigma(\cdot)$ are called *activations*. In general, with activation functions in place, it is no longer possible to collapse our MLP into a linear model:

$$\mathbf{H} = \sigma(\mathbf{X}\mathbf{W}^{(1)} + \mathbf{b}^{(1)}),$$
$$\mathbf{O} = \mathbf{H}\mathbf{W}^{(2)} + \mathbf{b}^{(2)}. \tag{5.1.3}$$

Since each row in $\mathbf{X}$ corresponds to an example in the minibatch, with some abuse of notation, we define the nonlinearity $\sigma$ to apply to its inputs in a rowwise fashion, i.e., one example at a time. Note that we used the same notation for softmax when we denoted a rowwise operation in Section 4.1.1. Quite frequently the activation functions we use apply not merely rowwise but elementwise. That means that after computing the linear portion of the layer, we can calculate each activation without looking at the values taken by the other hidden units.

To build more general MLPs, we can continue stacking such hidden layers, e.g., $\mathbf{H}^{(1)} = \sigma_1(\mathbf{X}\mathbf{W}^{(1)} + \mathbf{b}^{(1)})$ and $\mathbf{H}^{(2)} = \sigma_2(\mathbf{H}^{(1)}\mathbf{W}^{(2)} + \mathbf{b}^{(2)})$, one atop another, yielding ever more expressive models.

### Universal Approximators

We know that the brain is capable of very sophisticated statistical analysis. As such, it is worth asking, just *how powerful* a deep network could be. This question has been answered multiple times, e.g., in Cybenko (1989) in the context of MLPs, and in Micchelli (1984) in the context of reproducing kernel Hilbert spaces in a way that could be seen as radial basis function (RBF) networks with a single hidden layer. These (and related results) suggest that even with a single-hidden-layer network, given enough nodes (possibly absurdly many), and the right set of weights, we can model any function. Actually learning that function is the hard part, though. You might think of your neural network as being a bit like the C programming language. The language, like any other modern language, is capable of expressing any computable program. But actually coming up with a program that meets your specifications is the hard part.

Moreover, just because a single-hidden-layer network *can* learn any function does not mean that you should try to solve all of your problems with one. In fact, in this case kernel methods are way more effective, since they are capable of solving the problem *exactly* even in infinite dimensional spaces (Kimeldorf and Wahba, 1971, Schölkopf *et al.*, 2001). In fact, we can approximate many functions much more compactly by using deeper (rather than wider) networks (Simonyan and Zisserman, 2014). We will touch upon more rigorous arguments in subsequent chapters.

## 5.1.2 Activation Functions

Activation functions decide whether a neuron should be activated or not by calculating the weighted sum and further adding bias to it. They are differentiable operators for transforming input signals to outputs, while most of them add nonlinearity. Because activation functions are fundamental to deep learning, let's briefly survey some common ones.

### ReLU Function

The most popular choice, due to both simplicity of implementation and its good performance on a variety of predictive tasks, is the *rectified linear unit* (*ReLU*) (Nair and Hinton, 2010). ReLU provides a very simple nonlinear transformation. Given an element $x$, the function is defined as the maximum of that element and 0:

$$\operatorname{ReLU}(x) = \max(x, 0). \tag{5.1.4}$$

Informally, the ReLU function retains only positive elements and discards all negative elements by setting the corresponding activations to 0. To gain some intuition, we can plot the function. As you can see, the activation function is piecewise linear.

```
x = torch.arange(-8.0, 8.0, 0.1, requires_grad=True)
y = torch.relu(x)
d2l.plot(x.detach(), y.detach(), 'x', 'relu(x)', figsize=(5, 2.5))
```

When the input is negative, the derivative of the ReLU function is 0, and when the input is positive, the derivative of the ReLU function is 1. Note that the ReLU function is not differentiable when the input takes value precisely equal to 0. In these cases, we default to the left-hand-side derivative and say that the derivative is 0 when the input is 0. We can get away with this because the input may never actually be zero (mathematicians would say that it is nondifferentiable on a set of measure zero). There is an old adage that if subtle boundary conditions matter, we are probably doing (*real*) mathematics, not engineering. That conventional wisdom may apply here, or at least, the fact that we are not performing constrained optimization (Mangasarian, 1965, Rockafellar, 1970). We plot the derivative of the ReLU function below.

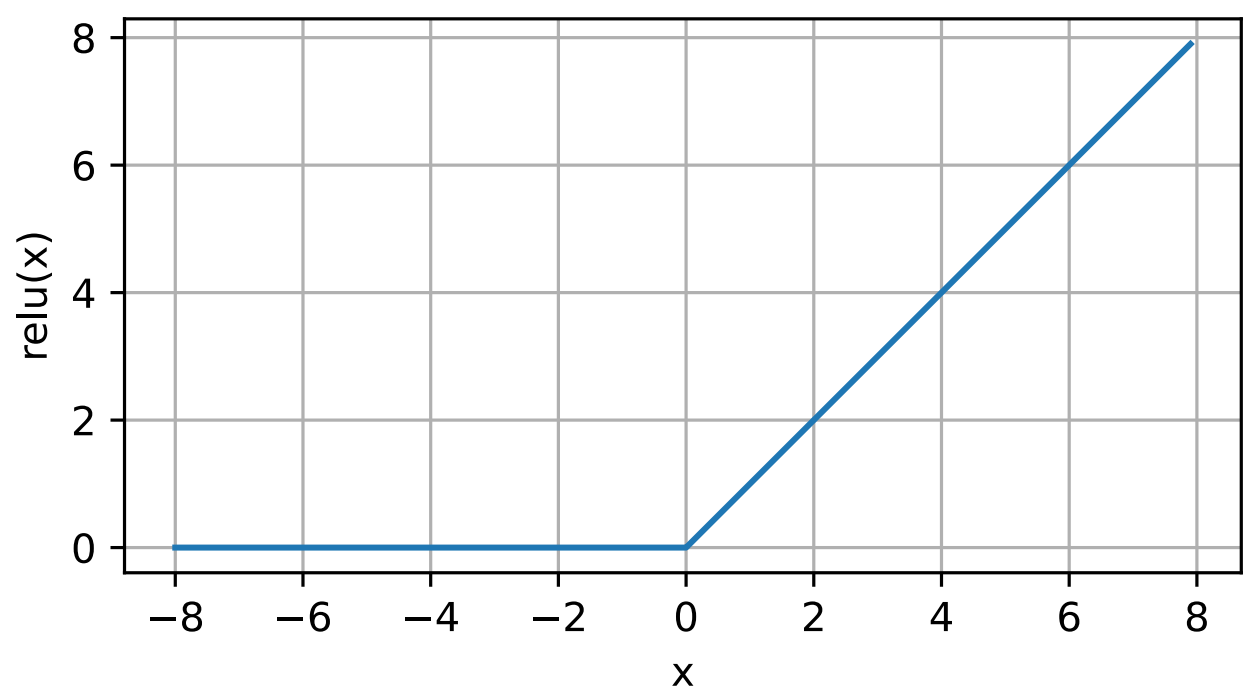

```
y.backward(torch.ones_like(x), retain_graph=True)
d2l.plot(x.detach(), x.grad, 'x', 'grad of relu', figsize=(5, 2.5))
```

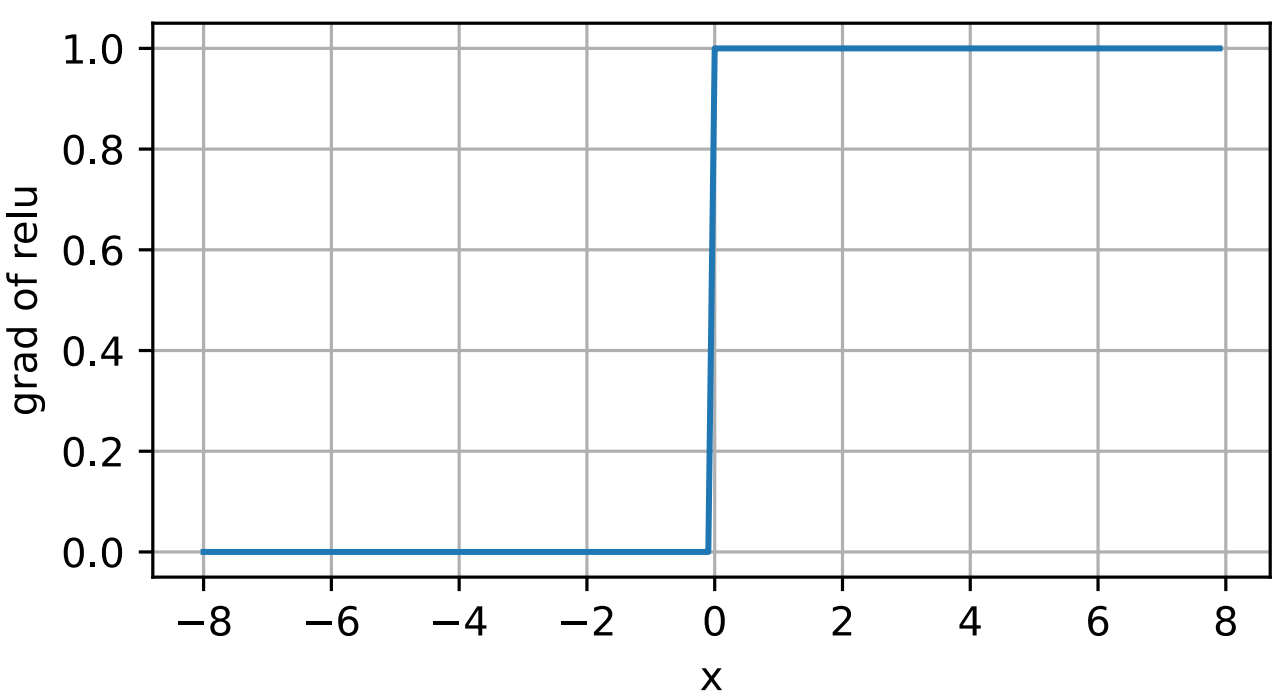

The reason for using ReLU is that its derivatives are particularly well behaved: either they vanish or they just let the argument through. This makes optimization better behaved and it mitigated the well-documented problem of vanishing gradients that plagued previous versions of neural networks (more on this later).

Note that there are many variants to the ReLU function, including the *parametrized ReLU* (*pReLU*) function (He *et al.*, 2015). This variation adds a linear term to ReLU, so some information still gets through, even when the argument is negative:

$$\mathrm{pReLU}(x) = \max(0, x) + \alpha \min(0, x). \tag{5.1.5}$$

### Sigmoid Function

The *sigmoid function* transforms those inputs whose values lie in the domain $\mathbb{R}$, to outputs that lie on the interval (0, 1). For that reason, the sigmoid is often called a *squashing function*: it squashes any input in the range (-inf, inf) to some value in the range (0, 1):

$$\text{sigmoid}(x) = \frac{1}{1 + \exp(-x)}. \tag{5.1.6}$$

In the earliest neural networks, scientists were interested in modeling biological neurons that either *fire* or *do not fire*. Thus the pioneers of this field, going all the way back to McCulloch and Pitts, the inventors of the artificial neuron, focused on thresholding units (McCulloch and Pitts, 1943). A thresholding activation takes value 0 when its input is below some threshold and value 1 when the input exceeds the threshold.

When attention shifted to gradient-based learning, the sigmoid function was a natural choice because it is a smooth, differentiable approximation to a thresholding unit. Sigmoids are still widely used as activation functions on the output units when we want to interpret the outputs as probabilities for binary classification problems: you can think of the sigmoid as a special case of the softmax. However, the sigmoid has largely been replaced by the simpler and more easily trainable ReLU for most use in hidden layers. Much of this has to do with the fact that the sigmoid poses challenges for optimization (LeCun *et al.*, 1998) since its gradient vanishes for large positive *and* negative arguments. This can lead to plateaus that are difficult to escape from. Nonetheless sigmoids are important. In later chapters (e.g., Section 10.1) on recurrent neural networks, we will describe architectures that leverage sigmoid units to control the flow of information across time.

Below, we plot the sigmoid function. Note that when the input is close to 0, the sigmoid function approaches a linear transformation.

```
y = torch.sigmoid(x)
d2l.plot(x.detach(), y.detach(), 'x', 'sigmoid(x)', figsize=(5, 2.5))
```

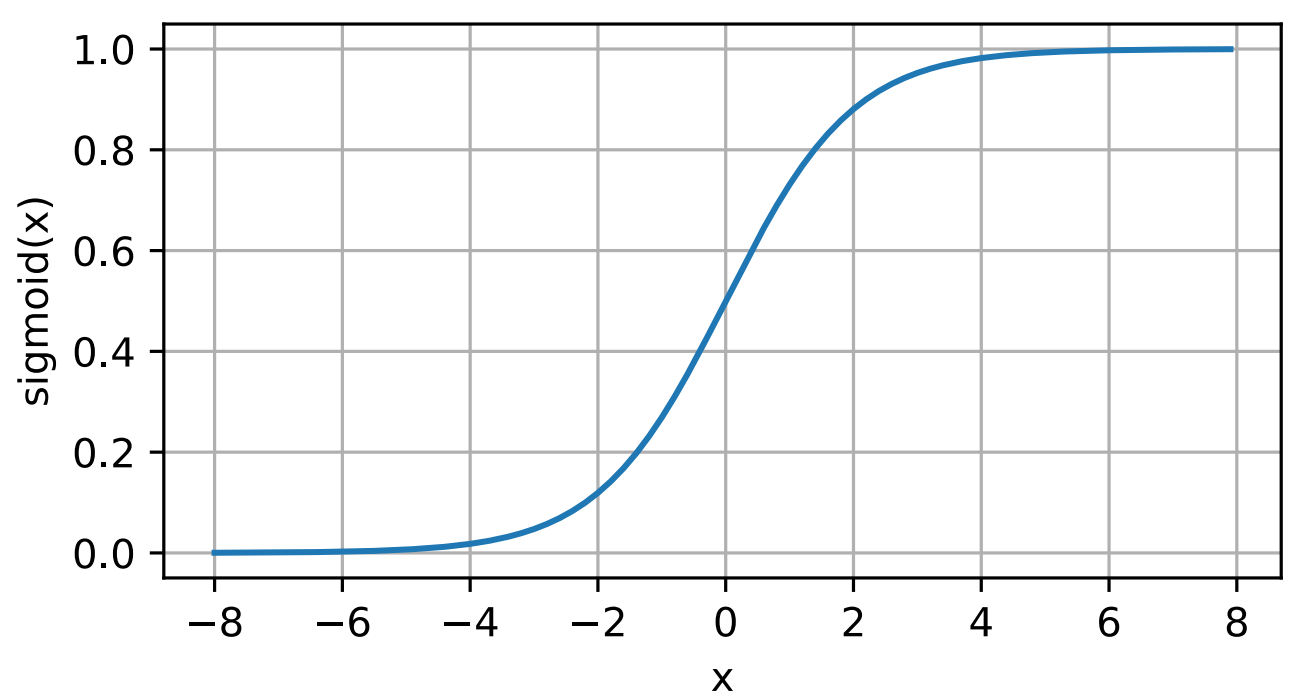

The derivative of the sigmoid function is given by the following equation:

$$\frac{d}{dx} \text{sigmoid}(x) = \frac{\exp(-x)}{(1 + \exp(-x))^2} = \text{sigmoid}(x) \left(1 - \text{sigmoid}(x)\right). \tag{5.1.7}$$

The derivative of the sigmoid function is plotted below. Note that when the input is 0, the derivative of the sigmoid function reaches a maximum of 0.25. As the input diverges from 0 in either direction, the derivative approaches 0.

```
# Clear out previous gradients
x.grad.data.zero_()
y.backward(torch.ones_like(x),retain_graph=True)
d2l.plot(x.detach(), x.grad, 'x', 'grad of sigmoid', figsize=(5, 2.5))
```

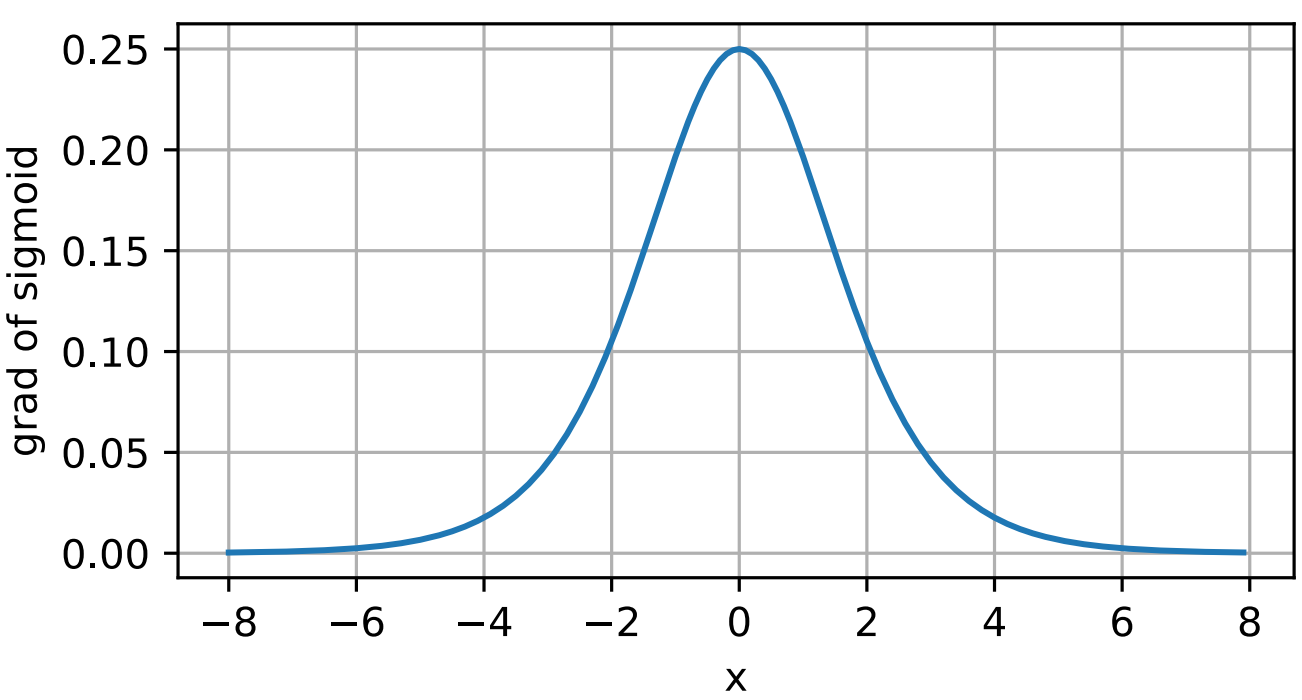

## Tanh Function

Like the sigmoid function, the tanh (hyperbolic tangent) function also squashes its inputs, transforming them into elements on the interval between $-1$ and $1$:

$$\tanh(x) = \frac{1 - \exp(-2x)}{1 + \exp(-2x)}. \tag{5.1.8}$$

We plot the tanh function below. Note that as input nears 0, the tanh function approaches a linear transformation. Although the shape of the function is similar to that of the sigmoid function, the tanh function exhibits point symmetry about the origin of the coordinate system (Kalman and Kwasny, 1992).

```
y = torch.tanh(x)
d2l.plot(x.detach(), y.detach(), 'x', 'tanh(x)', figsize=(5, 2.5))
```

The derivative of the tanh function is:

$$\frac{d}{dx} \tanh(x) = 1 - \tanh^2(x). \tag{5.1.9}$$

It is plotted below. As the input nears 0, the derivative of the tanh function approaches a maximum of 1. And as we saw with the sigmoid function, as input moves away from 0 in either direction, the derivative of the tanh function approaches 0.

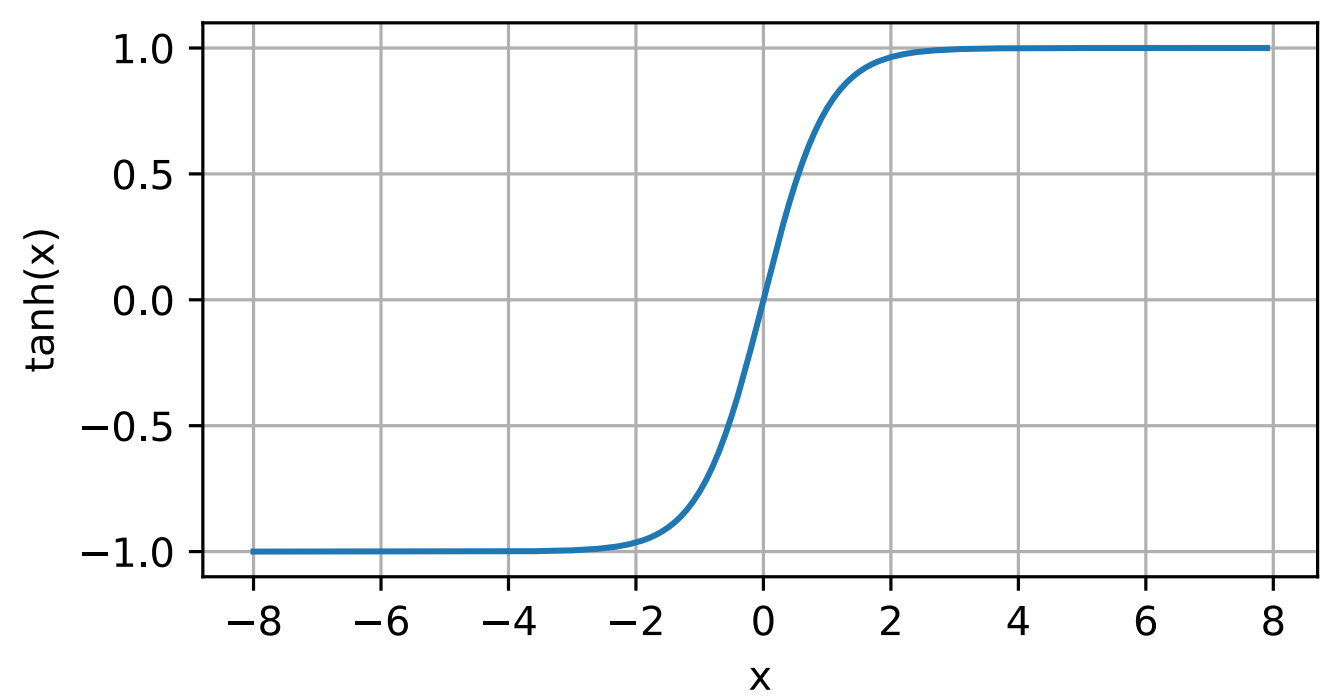

```
# Clear out previous gradients
x.grad.data.zero_()
y.backward(torch.ones_like(x),retain_graph=True)
d2l.plot(x.detach(), x.grad, 'x', 'grad of tanh', figsize=(5, 2.5))
```

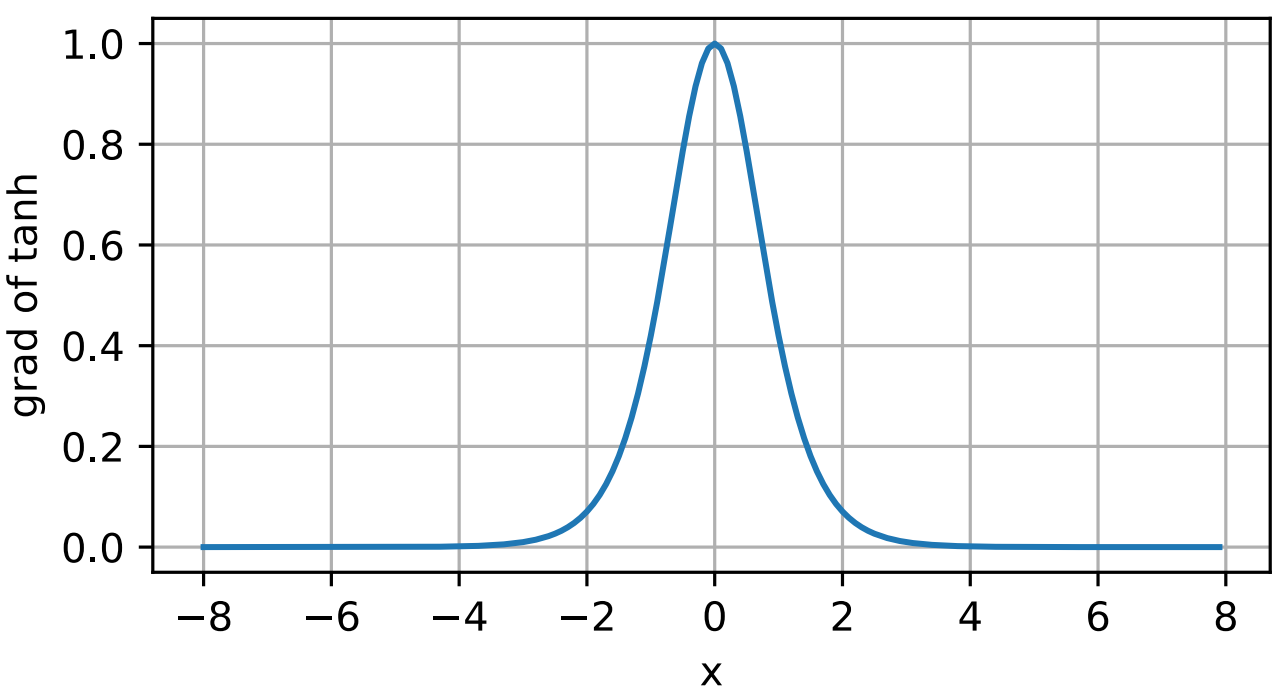

### 5.1.3 Summary and Discussion

We now know how to incorporate nonlinearities to build expressive multilayer neural network architectures. As a side note, your knowledge already puts you in command of a similar toolkit to a practitioner circa 1990. In some ways, you have an advantage over anyone working back then, because you can leverage powerful open-source deep learning frameworks to build models rapidly, using only a few lines of code. Previously, training these networks required researchers to code up layers and derivatives explicitly in C, Fortran, or even Lisp (in the case of LeNet).

A secondary benefit is that ReLU is significantly more amenable to optimization than the sigmoid or the tanh function. One could argue that this was one of the key innovations that helped the resurgence of deep learning over the past decade. Note, though, that research in activation functions has not stopped. For instance, the GELU (Gaussian error linear unit) activation function $x\Phi(x)$ by Hendrycks and Gimpel (2016) ($\Phi(x)$ is the standard Gaussian cumulative distribution function) and the Swish activation function $\sigma(x) = x \operatorname{sigmoid}(\beta x)$ as proposed in Ramachandran *et al.* (2017) can yield better accuracy in many cases.

### 5.1.4 Exercises

1. Show that adding layers to a *linear* deep network, i.e., a network without nonlinearity $\sigma$ can never increase the expressive power of the network. Give an example where it actively reduces it.
2. Compute the derivative of the pReLU activation function.
3. Compute the derivative of the Swish activation function $x \operatorname{sigmoid}(\beta x)$.
4. Show that an MLP using only ReLU (or pReLU) constructs a continuous piecewise linear function.
5. Sigmoid and tanh are very similar.
   1. Show that $\tanh(x) + 1 = 2 \operatorname{sigmoid}(2x)$.
   2. Prove that the function classes parametrized by both nonlinearities are identical. Hint: affine layers have bias terms, too.
6. Assume that we have a nonlinearity that applies to one minibatch at a time, such as the batch normalization (Ioffe and Szegedy, 2015). What kinds of problems do you expect this to cause?
7. Provide an example where the gradients vanish for the sigmoid activation function.

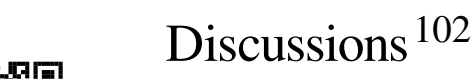

Discussions[102].

## 5.2 Implementation of Multilayer Perceptrons

Multilayer perceptrons (MLPs) are not much more complex to implement than simple linear models. The key conceptual difference is that we now concatenate multiple layers.

```
import torch
from torch import nn
from d2l import torch as d2l
```

### 5.2.1 Implementation from Scratch

Let's begin again by implementing such a network from scratch.

### Initializing Model Parameters

Recall that Fashion-MNIST contains 10 classes, and that each image consists of a $28 \times 28 = 784$ grid of grayscale pixel values. As before we will disregard the spatial structure among the pixels for now, so we can think of this as a classification dataset with 784 input features and 10 classes. To begin, we will implement an MLP with one hidden layer and 256 hidden units. Both the number of layers and their width are adjustable (they are considered hyperparameters). Typically, we choose the layer widths to be divisible by larger powers of 2. This is computationally efficient due to the way memory is allocated and addressed in hardware.

Again, we will represent our parameters with several tensors. Note that *for every layer*, we must keep track of one weight matrix and one bias vector. As always, we allocate memory for the gradients of the loss with respect to these parameters.

In the code below we use `nn.Parameter` to automatically register a class attribute as a parameter to be tracked by `autograd` (Section 2.5).

```
class MLPScratch(d2l.Classifier):
    def __init__(self, num_inputs, num_outputs, num_hiddens, lr, sigma=0.01):
        super().__init__()
        self.save_hyperparameters()
        self.W1 = nn.Parameter(torch.randn(num_inputs, num_hiddens) * sigma)
        self.b1 = nn.Parameter(torch.zeros(num_hiddens))
        self.W2 = nn.Parameter(torch.randn(num_hiddens, num_outputs) * sigma)
        self.b2 = nn.Parameter(torch.zeros(num_outputs))
```

### Model

To make sure we know how everything works, we will implement the ReLU activation ourselves rather than invoking the built-in `relu` function directly.

```
def relu(X):
    a = torch.zeros_like(X)
    return torch.max(X, a)
```

Since we are disregarding spatial structure, we `reshape` each two-dimensional image into a flat vector of length `num_inputs`. Finally, we implement our model with just a few lines of code. Since we use the framework built-in autograd this is all that it takes.

```
@d2l.add_to_class(MLPScratch)
def forward(self, X):
    X = X.reshape((-1, self.num_inputs))
    H = relu(torch.matmul(X, self.W1) + self.b1)
    return torch.matmul(H, self.W2) + self.b2
```

### Training

Fortunately, the training loop for MLPs is exactly the same as for softmax regression. We define the model, data, and trainer, then finally invoke the `fit` method on model and data.

```
model = MLPScratch(num_inputs=784, num_outputs=10, num_hiddens=256, lr=0.1)
data = d2l.FashionMNIST(batch_size=256)
trainer = d2l.Trainer(max_epochs=10)
trainer.fit(model, data)
```

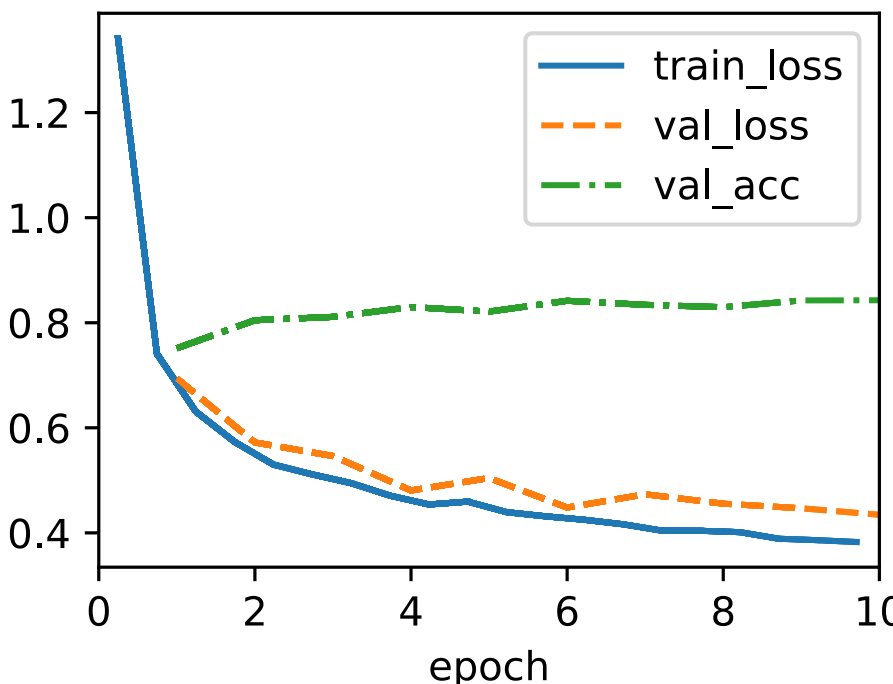

## 5.2.2 Concise Implementation

As you might expect, by relying on the high-level APIs, we can implement MLPs even more concisely.

### Model

Compared with our concise implementation of softmax regression implementation (Section 4.5), the only difference is that we add *two* fully connected layers where we previously added only *one*. The first is the hidden layer, the second is the output layer.

```
class MLP(d2l.Classifier):
    def __init__(self, num_outputs, num_hiddens, lr):
        super().__init__()
        self.save_hyperparameters()
        self.net = nn.Sequential(nn.Flatten(), nn.LazyLinear(num_hiddens),
                                 nn.ReLU(), nn.LazyLinear(num_outputs))
```

Previously, we defined `forward` methods for models to transform input using the model parameters. These operations are essentially a pipeline: you take an input and apply a transformation (e.g., matrix multiplication with weights followed by bias addition), then repetitively use the output of the current transformation as input to the next transformation. However, you may have noticed that no `forward` method is defined here. In fact, `MLP` inherits the `forward` method from the `Module` class (Section 3.2.2) to simply invoke `self.net(X)` (`X` is input), which is now defined as a sequence of transformations via the `Sequential` class. The `Sequential` class abstracts the forward process enabling us to focus on the transformations. We will further discuss how the `Sequential` class works in Section 6.1.2.

### Training

The training loop is exactly the same as when we implemented softmax regression. This modularity enables us to separate matters concerning the model architecture from orthogonal considerations.

```
model = MLP(num_outputs=10, num_hiddens=256, lr=0.1)
trainer.fit(model, data)
```

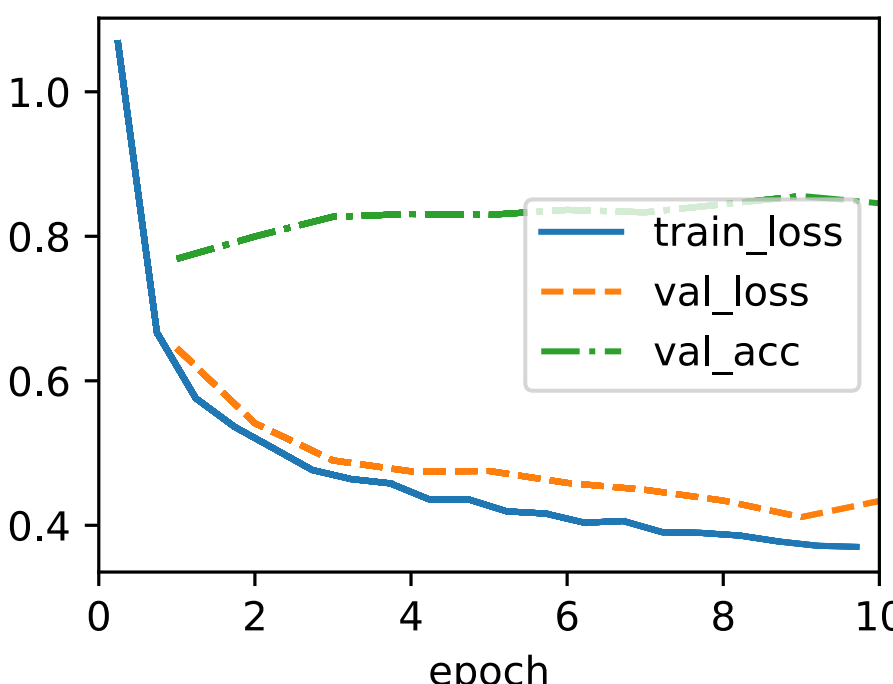

## 5.2.3 Summary

Now that we have more practice in designing deep networks, the step from a single to multiple layers of deep networks does not pose such a significant challenge any longer. In particular, we can reuse the training algorithm and data loader. Note, though, that implementing MLPs from scratch is nonetheless messy: naming and keeping track of the model parameters makes it difficult to extend models. For instance, imagine wanting to insert another layer between layers 42 and 43. This might now be layer 42b, unless we are willing to perform sequential renaming. Moreover, if we implement the network from scratch, it is much more difficult for the framework to perform meaningful performance optimizations.

Nonetheless, you have now reached the state of the art of the late 1980s when fully connected deep networks were the method of choice for neural network modeling. Our next conceptual step will be to consider images. Before we do so, we need to review a number of statistical basics and details on how to compute models efficiently.

### 5.2.4 Exercises

1. Change the number of hidden units `num_hiddens` and plot how its number affects the accuracy of the model. What is the best value of this hyperparameter?
2. Try adding a hidden layer to see how it affects the results.
3. Why is it a bad idea to insert a hidden layer with a single neuron? What could go wrong?
4. How does changing the learning rate alter your results? With all other parameters fixed, which learning rate gives you the best results? How does this relate to the number of epochs?
5. Let's optimize over all hyperparameters jointly, i.e., learning rate, number of epochs, number of hidden layers, and number of hidden units per layer.
    1. What is the best result you can get by optimizing over all of them?
    2. Why it is much more challenging to deal with multiple hyperparameters?
    3. Describe an efficient strategy for optimizing over multiple parameters jointly.
6. Compare the speed of the framework and the from-scratch implementation for a challenging problem. How does it change with the complexity of the network?
7. Measure the speed of tensor–matrix multiplications for well-aligned and misaligned matrices. For instance, test for matrices with dimension 1024, 1025, 1026, 1028, and 1032.
    1. How does this change between GPUs and CPUs?
    2. Determine the memory bus width of your CPU and GPU.
8. Try out different activation functions. Which one works best?
9. Is there a difference between weight initializations of the network? Does it matter?

Discussions[103].

## 5.3 Forward Propagation, Backward Propagation, and Computational Graphs

So far, we have trained our models with minibatch stochastic gradient descent. However, when we implemented the algorithm, we only worried about the calculations involved in *forward propagation* through the model. When it came time to calculate the gradients, we just invoked the backpropagation function provided by the deep learning framework.

The automatic calculation of gradients profoundly simplifies the implementation of deep learning algorithms. Before automatic differentiation, even small changes to complicated models required recalculating complicated derivatives by hand. Surprisingly often, academic papers had to allocate numerous pages to deriving update rules. While we must continue to rely on automatic differentiation so we can focus on the interesting parts, you ought to know how these gradients are calculated under the hood if you want to go beyond a shallow understanding of deep learning.

In this section, we take a deep dive into the details of *backward propagation* (more commonly called *backpropagation*). To convey some insight for both the techniques and their implementations, we rely on some basic mathematics and computational graphs. To start, we focus our exposition on a one-hidden-layer MLP with weight decay ($\ell_2$ regularization, to be described in subsequent chapters).

## 5.3.1 Forward Propagation

*Forward propagation* (or *forward pass*) refers to the calculation and storage of intermediate variables (including outputs) for a neural network in order from the input layer to the output layer. We now work step-by-step through the mechanics of a neural network with one hidden layer. This may seem tedious but in the eternal words of funk virtuoso James Brown, you must "pay the cost to be the boss".

For the sake of simplicity, let's assume that the input example is $\mathbf{x} \in \mathbb{R}^d$ and that our hidden layer does not include a bias term. Here the intermediate variable is:

$$\mathbf{z} = \mathbf{W}^{(1)} \mathbf{x}, \tag{5.3.1}$$

where $\mathbf{W}^{(1)} \in \mathbb{R}^{h \times d}$ is the weight parameter of the hidden layer. After running the intermediate variable $\mathbf{z} \in \mathbb{R}^h$ through the activation function $\phi$ we obtain our hidden activation vector of length $h$:

$$\mathbf{h} = \phi(\mathbf{z}). \tag{5.3.2}$$

The hidden layer output $\mathbf{h}$ is also an intermediate variable. Assuming that the parameters of the output layer possess only a weight of $\mathbf{W}^{(2)} \in \mathbb{R}^{q \times h}$, we can obtain an output layer variable with a vector of length $q$:

$$\mathbf{o} = \mathbf{W}^{(2)} \mathbf{h}. \tag{5.3.3}$$

Assuming that the loss function is $l$ and the example label is $y$, we can then calculate the loss term for a single data example,

$$L = l(\mathbf{o}, y). \tag{5.3.4}$$

As we will see the definition of $\ell_2$ regularization to be introduced later, given the hyperparameter $\lambda$, the regularization term is

$$s = \frac{\lambda}{2} \left(\|\mathbf{W}^{(1)}\|_\textrm{F}^2 + \|\mathbf{W}^{(2)}\|_\textrm{F}^2\right), \tag{5.3.5}$$

where the Frobenius norm of the matrix is simply the $\ell_2$ norm applied after flattening the matrix into a vector. Finally, the model's regularized loss on a given data example is:

$$J = L + s. \tag{5.3.6}$$

We refer to $J$ as the *objective function* in the following discussion.

### 5.3.2 Computational Graph of Forward Propagation

Plotting *computational graphs* helps us visualize the dependencies of operators and variables within the calculation. Fig. 5.3.1 contains the graph associated with the simple network described above, where squares denote variables and circles denote operators. The lower-left corner signifies the input and the upper-right corner is the output. Notice that the directions of the arrows (which illustrate data flow) are primarily rightward and upward.

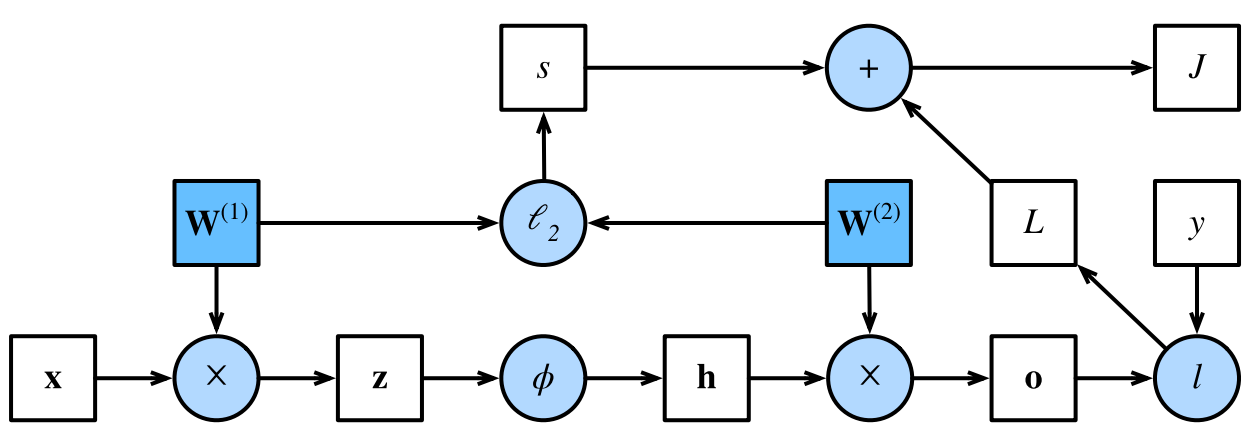

Fig. 5.3.1 Computational graph of forward propagation.

### 5.3.3 Backpropagation

*Backpropagation* refers to the method of calculating the gradient of neural network parameters. In short, the method traverses the network in reverse order, from the output to the input layer, according to the *chain rule* from calculus. The algorithm stores any intermediate variables (partial derivatives) required while calculating the gradient with respect to some parameters. Assume that we have functions $\mathsf{Y} = f(\mathsf{X})$ and $\mathsf{Z} = g(\mathsf{Y})$, in which the input and the output $\mathsf{X}, \mathsf{Y}, \mathsf{Z}$ are tensors of arbitrary shapes. By using the chain rule, we can compute the derivative of $\mathsf{Z}$ with respect to $\mathsf{X}$ via

$$\frac{\partial \mathsf{Z}}{\partial \mathsf{X}} = \text{prod}\left(\frac{\partial \mathsf{Z}}{\partial \mathsf{Y}}, \frac{\partial \mathsf{Y}}{\partial \mathsf{X}}\right). \tag{5.3.7}$$

Here we use the prod operator to multiply its arguments after the necessary operations, such as transposition and swapping input positions, have been carried out. For vectors, this is straightforward: it is simply matrix–matrix multiplication. For higher dimensional tensors, we use the appropriate counterpart. The operator prod hides all the notational overhead.

Recall that the parameters of the simple network with one hidden layer, whose computational graph is in Fig. 5.3.1, are $\mathbf{W}^{(1)}$ and $\mathbf{W}^{(2)}$. The objective of backpropagation is to calculate the gradients $\partial J/\partial \mathbf{W}^{(1)}$ and $\partial J/\partial \mathbf{W}^{(2)}$. To accomplish this, we apply the chain rule and calculate, in turn, the gradient of each intermediate variable and parameter. The order of calculations are reversed relative to those performed in forward propagation, since we need to start with the outcome of the computational graph and work our way towards the parameters. The first step is to calculate the gradients of the objective function $J = L + s$ with respect to the loss term $L$ and the regularization term $s$:

$$\frac{\partial J}{\partial L} = 1 \text{ and } \frac{\partial J}{\partial s} = 1. \tag{5.3.8}$$

Next, we compute the gradient of the objective function with respect to variable of the output layer $\mathbf{o}$ according to the chain rule:

$$\frac{\partial J}{\partial \mathbf{o}} = \text{prod}\left(\frac{\partial J}{\partial L}, \frac{\partial L}{\partial \mathbf{o}}\right) = \frac{\partial L}{\partial \mathbf{o}} \in \mathbb{R}^q. \tag{5.3.9}$$

Next, we calculate the gradients of the regularization term with respect to both parameters:

$$\frac{\partial s}{\partial \mathbf{W}^{(1)}} = \lambda \mathbf{W}^{(1)} \text{ and } \frac{\partial s}{\partial \mathbf{W}^{(2)}} = \lambda \mathbf{W}^{(2)}. \tag{5.3.10}$$

Now we are able to calculate the gradient $\partial J/\partial \mathbf{W}^{(2)} \in \mathbb{R}^{q \times h}$ of the model parameters closest to the output layer. Using the chain rule yields:

$$\frac{\partial J}{\partial \mathbf{W}^{(2)}} = \text{prod}\left(\frac{\partial J}{\partial \mathbf{o}}, \frac{\partial \mathbf{o}}{\partial \mathbf{W}^{(2)}}\right) + \text{prod}\left(\frac{\partial J}{\partial s}, \frac{\partial s}{\partial \mathbf{W}^{(2)}}\right) = \frac{\partial J}{\partial \mathbf{o}} \mathbf{h}^\top + \lambda \mathbf{W}^{(2)}. \tag{5.3.11}$$

To obtain the gradient with respect to $\mathbf{W}^{(1)}$ we need to continue backpropagation along the output layer to the hidden layer. The gradient with respect to the hidden layer output $\partial J/\partial \mathbf{h} \in \mathbb{R}^h$ is given by

$$\frac{\partial J}{\partial \mathbf{h}} = \text{prod}\left(\frac{\partial J}{\partial \mathbf{o}}, \frac{\partial \mathbf{o}}{\partial \mathbf{h}}\right) = {\mathbf{W}^{(2)}}^\top \frac{\partial J}{\partial \mathbf{o}}. \tag{5.3.12}$$

Since the activation function $\phi$ applies elementwise, calculating the gradient $\partial J/\partial \mathbf{z} \in \mathbb{R}^h$ of the intermediate variable $\mathbf{z}$ requires that we use the elementwise multiplication operator, which we denote by $\odot$:

$$\frac{\partial J}{\partial \mathbf{z}} = \text{prod}\left(\frac{\partial J}{\partial \mathbf{h}}, \frac{\partial \mathbf{h}}{\partial \mathbf{z}}\right) = \frac{\partial J}{\partial \mathbf{h}} \odot \phi'\left(\mathbf{z}\right). \tag{5.3.13}$$

Finally, we can obtain the gradient $\partial J/\partial \mathbf{W}^{(1)} \in \mathbb{R}^{h \times d}$ of the model parameters closest to the input layer. According to the chain rule, we get

$$\frac{\partial J}{\partial \mathbf{W}^{(1)}} = \text{prod}\left(\frac{\partial J}{\partial \mathbf{z}}, \frac{\partial \mathbf{z}}{\partial \mathbf{W}^{(1)}}\right) + \text{prod}\left(\frac{\partial J}{\partial s}, \frac{\partial s}{\partial \mathbf{W}^{(1)}}\right) = \frac{\partial J}{\partial \mathbf{z}} \mathbf{x}^\top + \lambda \mathbf{W}^{(1)}. \tag{5.3.14}$$

### 5.3.4 Training Neural Networks

When training neural networks, forward and backward propagation depend on each other. In particular, for forward propagation, we traverse the computational graph in the direction of dependencies and compute all the variables on its path. These are then used for backpropagation where the compute order on the graph is reversed.

Take the aforementioned simple network as an illustrative example. On the one hand, computing the regularization term (5.3.5) during forward propagation depends on the current values of model parameters $\mathbf{W}^{(1)}$ and $\mathbf{W}^{(2)}$. They are given by the optimization algorithm according to backpropagation in the most recent iteration. On the other hand, the gradient calculation for the parameter (5.3.11) during backpropagation depends on the current value of the hidden layer output $\mathbf{h}$, which is given by forward propagation.

Therefore when training neural networks, once model parameters are initialized, we alternate forward propagation with backpropagation, updating model parameters using gradients given by backpropagation. Note that backpropagation reuses the stored intermediate values from forward propagation to avoid duplicate calculations. One of the consequences is that we need to retain the intermediate values until backpropagation is complete. This is also one of the reasons why training requires significantly more memory than plain prediction. Besides, the size of such intermediate values is roughly proportional to the number of network layers and the batch size. Thus, training deeper networks using larger batch sizes more easily leads to *out-of-memory* errors.

### 5.3.5 Summary

Forward propagation sequentially calculates and stores intermediate variables within the computational graph defined by the neural network. It proceeds from the input to the output layer. Backpropagation sequentially calculates and stores the gradients of intermediate variables and parameters within the neural network in the reversed order. When training deep learning models, forward propagation and backpropagation are interdependent, and training requires significantly more memory than prediction.

### 5.3.6 Exercises

1. Assume that the inputs $\mathbf{X}$ to some scalar function $f$ are $n \times m$ matrices. What is the dimensionality of the gradient of $f$ with respect to $\mathbf{X}$?
2. Add a bias to the hidden layer of the model described in this section (you do not need to include bias in the regularization term).
    1. Draw the corresponding computational graph.
    2. Derive the forward and backward propagation equations.
3. Compute the memory footprint for training and prediction in the model described in this section.
4. Assume that you want to compute second derivatives. What happens to the computational graph? How long do you expect the calculation to take?
5. Assume that the computational graph is too large for your GPU.
    1. Can you partition it over more than one GPU?

2. What are the advantages and disadvantages over training on a smaller minibatch?

Discussions[104].

# 5.4 Numerical Stability and Initialization

Thus far, every model that we have implemented required that we initialize its parameters according to some pre-specified distribution. Until now, we took the initialization scheme for granted, glossing over the details of how these choices are made. You might have even gotten the impression that these choices are not especially important. On the contrary, the choice of initialization scheme plays a significant role in neural network learning, and it can be crucial for maintaining numerical stability. Moreover, these choices can be tied up in interesting ways with the choice of the nonlinear activation function. Which function we choose and how we initialize parameters can determine how quickly our optimization algorithm converges. Poor choices here can cause us to encounter exploding or vanishing gradients while training. In this section, we delve into these topics in greater detail and discuss some useful heuristics that you will find useful throughout your career in deep learning.

```
%matplotlib inline
import torch
from d2l import torch as d2l
```

## 5.4.1 Vanishing and Exploding Gradients

Consider a deep network with $L$ layers, input $\mathbf{x}$ and output $\mathbf{o}$. With each layer $l$ defined by a transformation $f_l$ parametrized by weights $\mathbf{W}^{(l)}$, whose hidden layer output is $\mathbf{h}^{(l)}$ (let $\mathbf{h}^{(0)} = \mathbf{x}$), our network can be expressed as:

$$\mathbf{h}^{(l)} = f_l(\mathbf{h}^{(l-1)}) \text{ and thus } \mathbf{o} = f_L \circ \cdots \circ f_1(\mathbf{x}). \tag{5.4.1}$$

If all the hidden layer output and the input are vectors, we can write the gradient of $\mathbf{o}$ with respect to any set of parameters $\mathbf{W}^{(l)}$ as follows:

$$\partial_{\mathbf{W}^{(l)}} \mathbf{o} = \underbrace{\partial_{\mathbf{h}^{(L-1)}} \mathbf{h}^{(L)}}_{ \mathbf{M}^{(L)} \stackrel{\textrm{def}}{=}} \cdots \underbrace{\partial_{\mathbf{h}^{(l)}} \mathbf{h}^{(l+1)}}_{ \mathbf{M}^{(l+1)} \stackrel{\textrm{def}}{=}} \underbrace{\partial_{\mathbf{W}^{(l)}} \mathbf{h}^{(l)}}_{ \mathbf{v}^{(l)} \stackrel{\textrm{def}}{=}}. \tag{5.4.2}$$

In other words, this gradient is the product of $L-l$ matrices $\mathbf{M}^{(L)} \cdots \mathbf{M}^{(l+1)}$ and the gradient vector $\mathbf{v}^{(l)}$. Thus we are susceptible to the same problems of numerical underflow that often crop up when multiplying together too many probabilities. When dealing with probabilities, a common trick is to switch into log-space, i.e., shifting pressure from the mantissa to the exponent of the numerical representation. Unfortunately, our problem above is more serious:

initially the matrices $\mathbf{M}^{(l)}$ may have a wide variety of eigenvalues. They might be small or large, and their product might be *very large* or *very small*.

The risks posed by unstable gradients go beyond numerical representation. Gradients of unpredictable magnitude also threaten the stability of our optimization algorithms. We may be facing parameter updates that are either (i) excessively large, destroying our model (the *exploding gradient* problem); or (ii) excessively small (the *vanishing gradient* problem), rendering learning impossible as parameters hardly move on each update.

### Vanishing Gradients

One frequent culprit causing the vanishing gradient problem is the choice of the activation function $\sigma$ that is appended following each layer's linear operations. Historically, the sigmoid function $1/(1 + \exp(-x))$ (introduced in Section 5.1) was popular because it resembles a thresholding function. Since early artificial neural networks were inspired by biological neural networks, the idea of neurons that fire either *fully* or *not at all* (like biological neurons) seemed appealing. Let's take a closer look at the sigmoid to see why it can cause vanishing gradients.

```
x = torch.arange(-8.0, 8.0, 0.1, requires_grad=True)
y = torch.sigmoid(x)
y.backward(torch.ones_like(x))

d2l.plot(x.detach().numpy(), [y.detach().numpy(), x.grad.numpy()],
         legend=['sigmoid', 'gradient'], figsize=(4.5, 2.5))
```

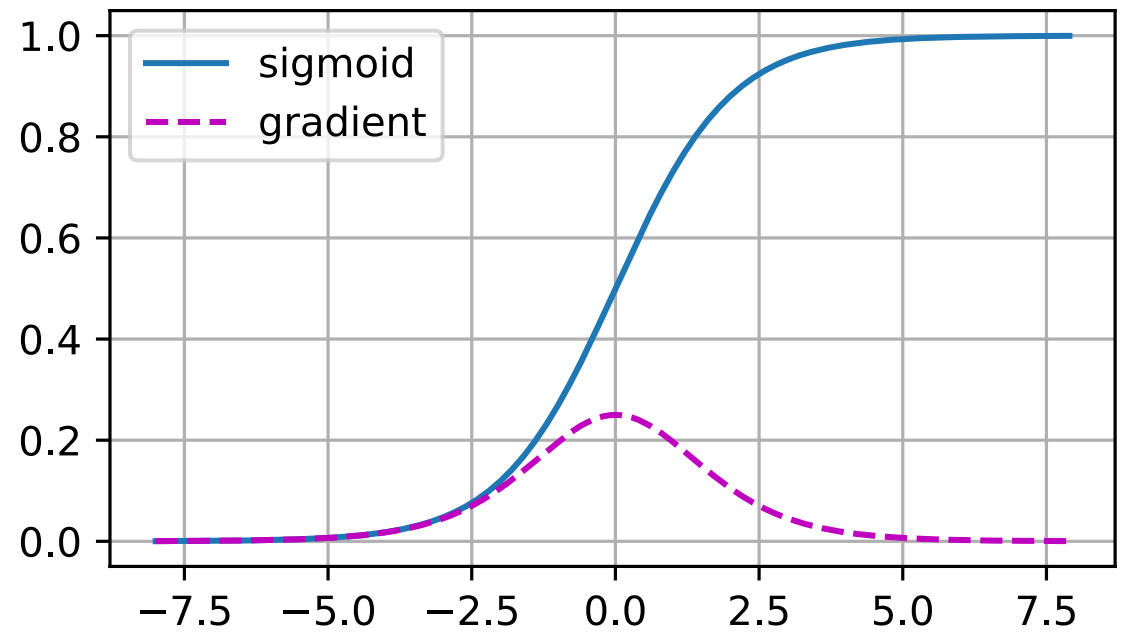

As you can see, the sigmoid's gradient vanishes both when its inputs are large and when they are small. Moreover, when backpropagating through many layers, unless we are in the Goldilocks zone, where the inputs to many of the sigmoids are close to zero, the gradients of the overall product may vanish. When our network boasts many layers, unless we are careful, the gradient will likely be cut off at some layer. Indeed, this problem used to plague deep network training. Consequently, ReLUs, which are more stable (but less neurally plausible), have emerged as the default choice for practitioners.

### Exploding Gradients

The opposite problem, when gradients explode, can be similarly vexing. To illustrate this a bit better, we draw 100 Gaussian random matrices and multiply them with some initial matrix. For the scale that we picked (the choice of the variance $\sigma^2 = 1$), the matrix product explodes. When this happens because of the initialization of a deep network, we have no chance of getting a gradient descent optimizer to converge.

```
M = torch.normal(0, 1, size=(4, 4))
print('a single matrix \n',M)
for i in range(100):
    M = M @ torch.normal(0, 1, size=(4, 4))
print('after multiplying 100 matrices\n', M)
```

```
a single matrix
 tensor([[ 1.7785,  1.4507,  1.3437,  0.6168],
        [-0.4228,  0.0714,  1.3975,  1.6048],
        [-1.0973,  0.2253,  0.3940, -0.2097],
        [-0.2455, -0.2039, -0.9628,  2.2473]])
after multiplying 100 matrices
 tensor([[-1.2044e+25,  2.8908e+22, -1.9161e+24, -4.9610e+24],
        [-8.1750e+24,  1.9621e+22, -1.3006e+24, -3.3674e+24],
        [ 2.8029e+24, -6.7275e+21,  4.4592e+23,  1.1545e+24],
        [-2.6804e+24,  6.4328e+21, -4.2643e+23, -1.1041e+24]])
```

### Breaking the Symmetry

Another problem in neural network design is the symmetry inherent in their parametrization. Assume that we have a simple MLP with one hidden layer and two units. In this case, we could permute the weights $\mathbf{W}^{(1)}$ of the first layer and likewise permute the weights of the output layer to obtain the same function. There is nothing special differentiating the first and second hidden units. In other words, we have permutation symmetry among the hidden units of each layer.

This is more than just a theoretical nuisance. Consider the aforementioned one-hidden-layer MLP with two hidden units. For illustration, suppose that the output layer transforms the two hidden units into only one output unit. Imagine what would happen if we initialized all the parameters of the hidden layer as $\mathbf{W}^{(1)} = c$ for some constant $c$. In this case, during forward propagation either hidden unit takes the same inputs and parameters producing the same activation which is fed to the output unit. During backpropagation, differentiating the output unit with respect to parameters $\mathbf{W}^{(1)}$ gives a gradient all of whose elements take the same value. Thus, after gradient-based iteration (e.g., minibatch stochastic gradient descent), all the elements of $\mathbf{W}^{(1)}$ still take the same value. Such iterations would never *break the symmetry* on their own and we might never be able to realize the network's expressive power. The hidden layer would behave as if it had only a single unit. Note that while minibatch stochastic gradient descent would not break this symmetry, dropout regularization (to be introduced later) would!

## 5.4.2 Parameter Initialization

One way of addressing—or at least mitigating—the issues raised above is through careful initialization. As we will see later, additional care during optimization and suitable regularization can further enhance stability.

### Default Initialization

In the previous sections, e.g., in Section 3.5, we used a normal distribution to initialize the values of our weights. If we do not specify the initialization method, the framework will use a default random initialization method, which often works well in practice for moderate problem sizes.

### Xavier Initialization

Let's look at the scale distribution of an output $o_i$ for some fully connected layer *without nonlinearities*. With $n_\textrm{in}$ inputs $x_j$ and their associated weights $w_{ij}$ for this layer, an output is given by

$$o_{i} = \sum_{j=1}^{n_\textrm{in}} w_{ij} x_j. \tag{5.4.3}$$

The weights $w_{ij}$ are all drawn independently from the same distribution. Furthermore, let's assume that this distribution has zero mean and variance $\sigma^2$. Note that this does not mean that the distribution has to be Gaussian, just that the mean and variance need to exist. For now, let's assume that the inputs to the layer $x_j$ also have zero mean and variance $\gamma^2$ and that they are independent of $w_{ij}$ and independent of each other. In this case, we can compute the mean of $o_i$:

$$\begin{aligned}
E[o_i] & = \sum_{j=1}^{n_\textrm{in}} E[w_{ij} x_j] \\
&= \sum_{j=1}^{n_\textrm{in}} E[w_{ij}] E[x_j] \\
&= 0,
\end{aligned} \tag{5.4.4}$$

and the variance:

$$\begin{aligned}\mathrm{Var}[o_i] &= E[o_i^2] - (E[o_i])^2 \\ &= \sum_{j=1}^{n_\mathrm{in}} E[w_{ij}^2 x_j^2] - 0 \\ &= \sum_{j=1}^{n_\mathrm{in}} E[w_{ij}^2] E[x_j^2] \\ &= n_\mathrm{in} \sigma^2 \gamma^2.\end{aligned} \tag{5.4.5}$$

One way to keep the variance fixed is to set $n_\mathrm{in}\sigma^2 = 1$. Now consider backpropagation. There we face a similar problem, albeit with gradients being propagated from the layers closer to the output. Using the same reasoning as for forward propagation, we see that the gradients' variance can blow up unless $n_\mathrm{out}\sigma^2 = 1$, where $n_\mathrm{out}$ is the number of outputs of this layer. This leaves us in a dilemma: we cannot possibly satisfy both conditions simultaneously. Instead, we simply try to satisfy:

$$\frac{1}{2} (n_\mathrm{in} + n_\mathrm{out}) \sigma^2 = 1 \text{ or equivalently } \sigma = \sqrt{\frac{2}{n_\mathrm{in} + n_\mathrm{out}}}. \tag{5.4.6}$$

This is the reasoning underlying the now-standard and practically beneficial *Xavier initialization*, named after the first author of its creators (Glorot and Bengio, 2010). Typically, the Xavier initialization samples weights from a Gaussian distribution with zero mean and variance $\sigma^2 = \frac{2}{n_\mathrm{in} + n_\mathrm{out}}$. We can also adapt this to choose the variance when sampling weights from a uniform distribution. Note that the uniform distribution $U(-a, a)$ has variance $\frac{a^2}{3}$. Plugging $\frac{a^2}{3}$ into our condition on $\sigma^2$ prompts us to initialize according to

$$U\left(-\sqrt{\frac{6}{n_\mathrm{in} + n_\mathrm{out}}}, \sqrt{\frac{6}{n_\mathrm{in} + n_\mathrm{out}}}\right). \tag{5.4.7}$$

Though the assumption for nonexistence of nonlinearities in the above mathematical reasoning can be easily violated in neural networks, the Xavier initialization method turns out to work well in practice.

## Beyond

The reasoning above barely scratches the surface of modern approaches to parameter initialization. A deep learning framework often implements over a dozen different heuristics. Moreover, parameter initialization continues to be a hot area of fundamental research in deep learning. Among these are heuristics specialized for tied (shared) parameters, super-resolution, sequence models, and other situations. For instance, Xiao *et al.* (2018) demonstrated the possibility of training 10,000-layer neural networks without architectural tricks by using a carefully-designed initialization method.

If the topic interests you we suggest a deep dive into this module's offerings, reading the papers that proposed and analyzed each heuristic, and then exploring the latest publications on the topic. Perhaps you will stumble across or even invent a clever idea and contribute an implementation to deep learning frameworks.

### 5.4.3 Summary

Vanishing and exploding gradients are common issues in deep networks. Great care in parameter initialization is required to ensure that gradients and parameters remain well controlled. Initialization heuristics are needed to ensure that the initial gradients are neither too large nor too small. Random initialization is key to ensuring that symmetry is broken before optimization. Xavier initialization suggests that, for each layer, variance of any output is not affected by the number of inputs, and variance of any gradient is not affected by the number of outputs. ReLU activation functions mitigate the vanishing gradient problem. This can accelerate convergence.

### 5.4.4 Exercises

1. Can you design other cases where a neural network might exhibit symmetry that needs breaking, besides the permutation symmetry in an MLP's layers?
2. Can we initialize all weight parameters in linear regression or in softmax regression to the same value?
3. Look up analytic bounds on the eigenvalues of the product of two matrices. What does this tell you about ensuring that gradients are well conditioned?
4. If we know that some terms diverge, can we fix this after the fact? Look at the paper on layerwise adaptive rate scaling for inspiration (You *et al.*, 2017).

Discussions[105].

## 5.5 Generalization in Deep Learning

In Chapter 3 and Chapter 4, we tackled regression and classification problems by fitting linear models to training data. In both cases, we provided practical algorithms for finding the parameters that maximized the likelihood of the observed training labels. And then, towards the end of each chapter, we recalled that fitting the training data was only an intermediate goal. Our real quest all along was to discover *general patterns* on the basis of which we can make accurate predictions even on new examples drawn from the same underlying population. Machine learning researchers are *consumers* of optimization algorithms. Sometimes, we must even develop new optimization algorithms. But at the end of the day, optimization is merely a means to an end. At its core, machine learning is a statistical discipline and we wish to optimize training loss only insofar as some statistical principle (known or unknown) leads the resulting models to generalize beyond the training set.

On the bright side, it turns out that deep neural networks trained by stochastic gradient descent generalize remarkably well across myriad prediction problems, spanning computer vision; natural language processing; time series data; recommender systems; electronic health records; protein folding; value function approximation in video games and board games; and numerous other domains. On the downside, if you were looking for a straightforward account of either the optimization story (why we can fit them to training data) or the generalization story (why the resulting models generalize to unseen examples), then you might want to pour yourself a drink. While our procedures for optimizing linear models and the statistical properties of the solutions are both described well by a comprehensive body of theory, our understanding of deep learning still resembles the wild west on both fronts.

Both the theory and practice of deep learning are rapidly evolving, with theorists adopting new strategies to explain what's going on, even as practitioners continue to innovate at a blistering pace, building arsenals of heuristics for training deep networks and a body of intuitions and folk knowledge that provide guidance for deciding which techniques to apply in which situations.

The summary of the present moment is that the theory of deep learning has produced promising lines of attack and scattered fascinating results, but still appears far from a comprehensive account of both (i) why we are able to optimize neural networks and (ii) how models learned by gradient descent manage to generalize so well, even on high-dimensional tasks. However, in practice, (i) is seldom a problem (we can always find parameters that will fit all of our training data) and thus understanding generalization is far the bigger problem. On the other hand, even absent the comfort of a coherent scientific theory, practitioners have developed a large collection of techniques that may help you to produce models that generalize well in practice. While no pithy summary can possibly do justice to the vast topic of generalization in deep learning, and while the overall state of research is far from resolved, we hope, in this section, to present a broad overview of the state of research and practice.

### 5.5.1 Revisiting Overfitting and Regularization

According to the "no free lunch" theorem of Wolpert and Macready (1995), any learning algorithm generalizes better on data with certain distributions, and worse with other distributions. Thus, given a finite training set, a model relies on certain assumptions: to achieve human-level performance it may be useful to identify *inductive biases* that reflect how humans think about the world. Such inductive biases show preferences for solutions with certain properties. For example, a deep MLP has an inductive bias towards building up a complicated function by the composition of simpler functions.

With machine learning models encoding inductive biases, our approach to training them typically consists of two phases: (i) fit the training data; and (ii) estimate the *generalization error* (the true error on the underlying population) by evaluating the model on holdout data. The difference between our fit on the training data and our fit on the test data is called the *gen-* *eralization gap* and when this is large, we say that our models *overfit* to the training data. In extreme cases of overfitting, we might exactly fit the training data, even when the test error remains significant. And in the classical view, the interpretation is that our models are too complex, requiring that we either shrink the number of features, the number of nonzero parameters learned, or the size of the parameters as quantified. Recall the plot of model complexity compared with loss (Fig. 3.6.1) from Section 3.6.

However deep learning complicates this picture in counterintuitive ways. First, for classification problems, our models are typically expressive enough to perfectly fit every training example, even in datasets consisting of millions (Zhang *et al.*, 2021). In the classical picture, we might think that this setting lies on the far right extreme of the model complexity axis, and that any improvements in generalization error must come by way of regularization, either by reducing the complexity of the model class, or by applying a penalty, severely constraining the set of values that our parameters might take. But that is where things start to get weird.

Strangely, for many deep learning tasks (e.g., image recognition and text classification) we are typically choosing among model architectures, all of which can achieve arbitrarily low training loss (and zero training error). Because all models under consideration achieve zero training error, *the only avenue for further gains is to reduce overfitting*. Even stranger, it is often the case that despite fitting the training data perfectly, we can actually *reduce the generalization error* further by making the model *even more expressive*, e.g., adding layers, nodes, or training for a larger number of epochs. Stranger yet, the pattern relating the generalization gap to the *complexity* of the model (as captured, for example, in the depth or width of the networks) can be non-monotonic, with greater complexity hurting at first but subsequently helping in a so-called "double-descent" pattern (Nakkiran *et al.*, 2021). Thus the deep learning practitioner possesses a bag of tricks, some of which seemingly restrict the model in some fashion and others that seemingly make it even more expressive, and all of which, in some sense, are applied to mitigate overfitting.

Complicating things even further, while the guarantees provided by classical learning theory can be conservative even for classical models, they appear powerless to explain why it is that deep neural networks generalize in the first place. Because deep neural networks are capable of fitting arbitrary labels even for large datasets, and despite the use of familiar methods such as $\ell_2$ regularization, traditional complexity-based generalization bounds, e.g., those based on the VC dimension or Rademacher complexity of a hypothesis class cannot explain why neural networks generalize.

## 5.5.2 Inspiration from Nonparametrics

Approaching deep learning for the first time, it is tempting to think of them as parametric models. After all, the models *do* have millions of parameters. When we update the models, we update their parameters. When we save the models, we write their parameters to disk. However, mathematics and computer science are riddled with counterintuitive changes of perspective, and surprising isomorphisms between seemingly different problems. While neural networks clearly *have* parameters, in some ways it can be more fruitful to think of them as behaving like nonparametric models. So what precisely makes a model nonparametric? While the name covers a diverse set of approaches, one common theme is that nonparametric methods tend to have a level of complexity that grows as the amount of available data grows.

Perhaps the simplest example of a nonparametric model is the $k$-nearest neighbor algorithm (we will cover more nonparametric models later, for example in Section 11.2). Here, at training time, the learner simply memorizes the dataset. Then, at prediction time, when confronted with a new point $\mathbf{x}$, the learner looks up the $k$ nearest neighbors (the $k$ points $\mathbf{x}'_i$ that minimize some distance $d(\mathbf{x}, \mathbf{x}'_i)$). When $k = 1$, this algorithm is called 1-nearest neighbors, and the algorithm will always achieve a training error of zero. That however, does not mean that the algorithm will not generalize. In fact, it turns out that under some mild conditions, the 1-nearest neighbor algorithm is consistent (eventually converging to the optimal predictor).

Note that 1-nearest neighbor requires that we specify some distance function $d$, or equivalently, that we specify some vector-valued basis function $\phi(\mathbf{x})$ for featurizing our data. For any choice of the distance metric, we will achieve zero training error and eventually reach an optimal predictor, but different distance metrics $d$ encode different inductive biases and with a finite amount of available data will yield different predictors. Different choices of the distance metric $d$ represent different assumptions about the underlying patterns and the performance of the different predictors will depend on how compatible the assumptions are with the observed data.

In a sense, because neural networks are over-parametrized, possessing many more parameters than are needed to fit the training data, they tend to *interpolate* the training data (fitting it perfectly) and thus behave, in some ways, more like nonparametric models. More recent theoretical research has established deep connection between large neural networks and nonparametric methods, notably kernel methods. In particular, Jacot *et al.* (2018) demonstrated that in the limit, as multilayer perceptrons with randomly initialized weights grow infinitely wide, they become equivalent to (nonparametric) kernel methods for a specific choice of the kernel function (essentially, a distance function), which they call the neural tangent kernel. While current neural tangent kernel models may not fully explain the behavior of modern deep networks, their success as an analytical tool underscores the usefulness of nonparametric modeling for understanding the behavior of over-parametrized deep networks.

### 5.5.3 Early Stopping

While deep neural networks are capable of fitting arbitrary labels, even when labels are assigned incorrectly or randomly (Zhang *et al.*, 2021), this capability only emerges over many iterations of training. A new line of work (Rolnick *et al.*, 2017) has revealed that in the setting of label noise, neural networks tend to fit cleanly labeled data first and only subsequently to interpolate the mislabeled data. Moreover, it has been established that this phenomenon translates directly into a guarantee on generalization: whenever a model has fitted the cleanly labeled data but not randomly labeled examples included in the training set, it has in fact generalized (Garg *et al.*, 2021).

Together these findings help to motivate *early stopping*, a classic technique for regularizing deep neural networks. Here, rather than directly constraining the values of the weights, one constrains the number of epochs of training. The most common way to determine the stopping criterion is to monitor validation error throughout training (typically by checking once after each epoch) and to cut off training when the validation error has not decreased by more than some small amount $\epsilon$ for some number of epochs. This is sometimes called a *patience criterion*. As well as the potential to lead to better generalization in the setting of noisy labels, another benefit of early stopping is the time saved. Once the patience criterion is met, one can terminate training. For large models that might require days of training simultaneously across eight or more GPUs, well-tuned early stopping can save researchers days of time and can save their employers many thousands of dollars.

Notably, when there is no label noise and datasets are *realizable* (the classes are truly separable, e.g., distinguishing cats from dogs), early stopping tends not to lead to significant improvements in generalization. On the other hand, when there is label noise, or intrinsic variability in the label (e.g., predicting mortality among patients), early stopping is crucial. Training models until they interpolate noisy data is typically a bad idea.

### 5.5.4 Classical Regularization Methods for Deep Networks

In Chapter 3, we described several classical regularization techniques for constraining the complexity of our models. In particular, Section 3.7 introduced a method called weight decay, which consists of adding a regularization term to the loss function in order to penalize large values of the weights. Depending on which weight norm is penalized this technique is known either as ridge regularization (for $\ell_2$ penalty) or lasso regularization (for an $\ell_1$ penalty). In the classical analysis of these regularizers, they are considered as sufficiently restrictive on the values that the weights can take to prevent the model from fitting arbitrary labels.

In deep learning implementations, weight decay remains a popular tool. However, researchers have noted that typical strengths of $\ell_2$ regularization are insufficient to prevent the networks from interpolating the data (Zhang *et al.*, 2021) and thus the benefits if interpreted as regularization might only make sense in combination with the early stopping criterion. Absent early stopping, it is possible that just like the number of layers or number of nodes (in deep learning) or the distance metric (in 1-nearest neighbor), these methods may lead to better generalization not because they meaningfully constrain the power of the neural network but rather because they somehow encode inductive biases that are better compatible with the patterns found in datasets of interests. Thus, classical regularizers remain popular in deep learning implementations, even if the theoretical rationale for their efficacy may be radically different.

Notably, deep learning researchers have also built on techniques first popularized in classical regularization contexts, such as adding noise to model inputs. In the next section we will introduce the famous dropout technique (invented by Srivastava *et al.* (2014)), which has become a mainstay of deep learning, even as the theoretical basis for its efficacy remains similarly mysterious.

### 5.5.5 Summary

Unlike classical linear models, which tend to have fewer parameters than examples, deep networks tend to be over-parametrized, and for most tasks are capable of perfectly fitting the training set. This *interpolation regime* challenges many hard fast-held intuitions. Functionally, neural networks look like parametric models. But thinking of them as nonparametric models can sometimes be a more reliable source of intuition. Because it is often the case that all deep networks under consideration are capable of fitting all of the training labels, nearly all gains must come by mitigating overfitting (closing the *generalization gap*). Paradoxically, the interventions that reduce the generalization gap sometimes appear to increase model complexity and at other times appear to decrease complexity. However, these methods seldom decrease complexity sufficiently for classical theory to explain the generalization of deep networks, and *why certain choices lead to improved generalization* remains for the most part a massive open question despite the concerted efforts of many brilliant researchers.

### 5.5.6 Exercises

1. In what sense do traditional complexity-based measures fail to account for generalization of deep neural networks?
2. Why might *early stopping* be considered a regularization technique?
3. How do researchers typically determine the stopping criterion?
4. What important factor seems to differentiate cases when early stopping leads to big improvements in generalization?
5. Beyond generalization, describe another benefit of early stopping.

Discussions[106].

## 5.6 Dropout

Let's think briefly about what we expect from a good predictive model. We want it to peform well on unseen data. Classical generalization theory suggests that to close the gap between train and test performance, we should aim for a simple model. Simplicity can come in the form of a small number of dimensions. We explored this when discussing the monomial basis functions of linear models in Section 3.6. Additionally, as we saw when discussing weight decay ($\ell_2$ regularization) in Section 3.7, the (inverse) norm of the parameters also represents a useful measure of simplicity. Another useful notion of simplicity is smoothness, i.e., that the function should not be sensitive to small changes to its inputs. For instance, when we classify images, we would expect that adding some random noise to the pixels should be mostly harmless.

Bishop (1995) formalized this idea when he proved that training with input noise is equivalent to Tikhonov regularization. This work drew a clear mathematical connection between the requirement that a function be smooth (and thus simple), and the requirement that it be resilient to perturbations in the input.

Then, Srivastava *et al.* (2014) developed a clever idea for how to apply Bishop's idea to the internal layers of a network, too. Their idea, called *dropout*, involves injecting noise while computing each internal layer during forward propagation, and it has become a standard technique for training neural networks. The method is called *dropout* because we literally *drop out* some neurons during training. Throughout training, on each iteration, standard dropout consists of zeroing out some fraction of the nodes in each layer before calculating the subsequent layer.

To be clear, we are imposing our own narrative with the link to Bishop. The original paper on dropout offers intuition through a surprising analogy to sexual reproduction. The authors argue that neural network overfitting is characterized by a state in which each layer relies on a specific pattern of activations in the previous layer, calling this condition *co-adaptation*. Dropout, they claim, breaks up co-adaptation just as sexual reproduction is argued to break up co-adapted genes. While such an justification of this theory is certainly up for debate, the dropout technique itself has proved enduring, and various forms of dropout are implemented in most deep learning libraries.

The key challenge is how to inject this noise. One idea is to inject it in an *unbiased* manner so that the expected value of each layer—while fixing the others—equals the value it would have taken absent noise. In Bishop's work, he added Gaussian noise to the inputs to a linear model. At each training iteration, he added noise sampled from a distribution with mean zero $\epsilon \sim \mathcal{N}(0, \sigma^2)$ to the input $\mathbf{x}$, yielding a perturbed point $\mathbf{x}' = \mathbf{x} + \epsilon$. In expectation, $E[\mathbf{x}'] = \mathbf{x}$.

In standard dropout regularization, one zeros out some fraction of the nodes in each layer and then *debiases* each layer by normalizing by the fraction of nodes that were retained (not dropped out). In other words, with *dropout probability* $p$, each intermediate activation $h$ is replaced by a random variable $h'$ as follows:

$$h' = \begin{cases} 0 & \text{with probability } p \\ \frac{h}{1-p} & \text{otherwise} \end{cases} \tag{5.6.1}$$

By design, the expectation remains unchanged, i.e., $E[h'] = h$.

```
import torch
from torch import nn
from d2l import torch as d2l
```

## 5.6.1 Dropout in Practice

Recall the MLP with a hidden layer and five hidden units from Fig. 5.1.1. When we apply dropout to a hidden layer, zeroing out each hidden unit with probability $p$, the result can be viewed as a network containing only a subset of the original neurons. In Fig. 5.6.1, $h_2$ and $h_5$ are removed. Consequently, the calculation of the outputs no longer depends on $h_2$ or $h_5$ and their respective gradient also vanishes when performing backpropagation. In this way, the calculation of the output layer cannot be overly dependent on any one element of $h_1, \ldots, h_5$.

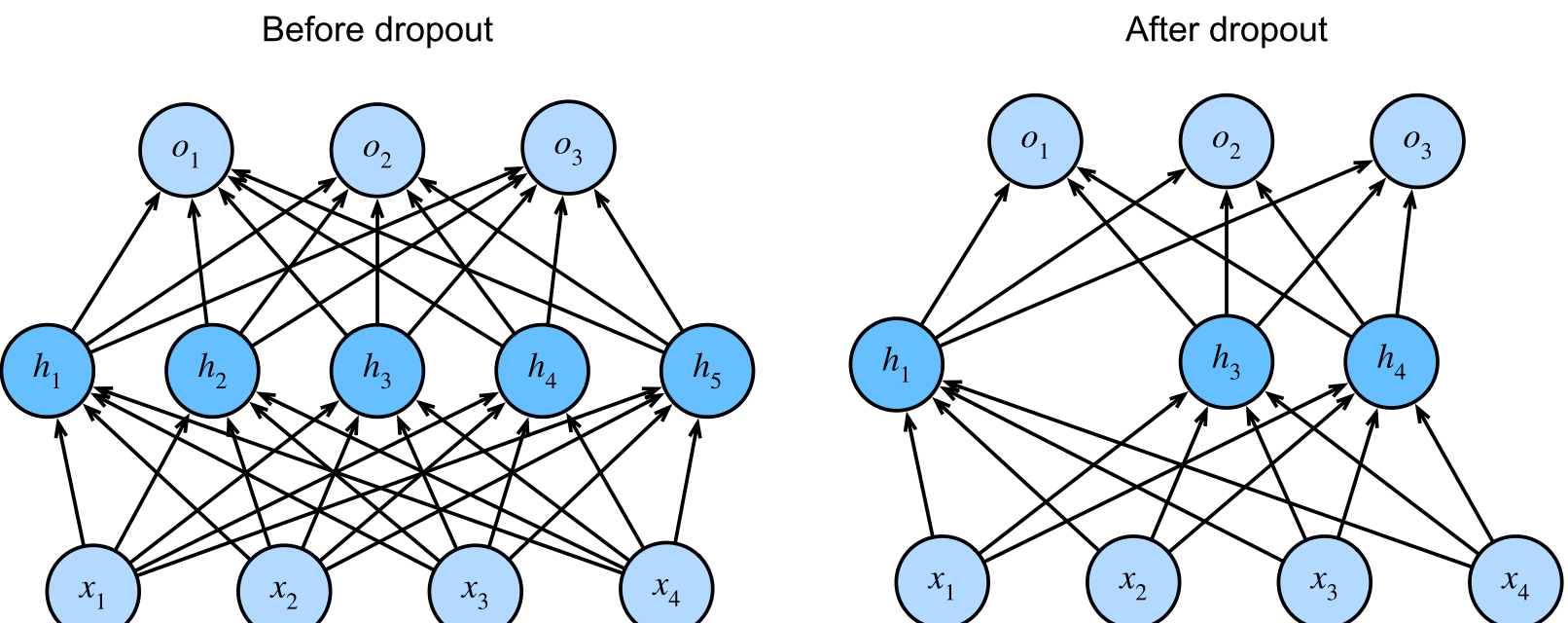

Fig. 5.6.1 MLP before and after dropout.

Typically, we disable dropout at test time. Given a trained model and a new example, we do not drop out any nodes and thus do not need to normalize. However, there are some exceptions: some researchers use dropout at test time as a heuristic for estimating the *uncertainty* of neural network predictions: if the predictions agree across many different dropout outputs, then we might say that the network is more confident.

## 5.6.2 Implementation from Scratch

To implement the dropout function for a single layer, we must draw as many samples from a Bernoulli (binary) random variable as our layer has dimensions, where the random variable takes value $1$ (keep) with probability $1 - p$ and $0$ (drop) with probability $p$. One easy way to implement this is to first draw samples from the uniform distribution $U[0, 1]$. Then we can keep those nodes for which the corresponding sample is greater than $p$, dropping the rest.

In the following code, we implement a `dropout_layer` function that drops out the elements in the tensor input `X` with probability `dropout`, rescaling the remainder as described above: dividing the survivors by `1.0-dropout`.

```
def dropout_layer(X, dropout):
    assert 0 <= dropout <= 1
    if dropout == 1: return torch.zeros_like(X)
    mask = (torch.rand(X.shape) > dropout).float()
    return mask * X / (1.0 - dropout)
```

We can test out the `dropout_layer` function on a few examples. In the following lines of code, we pass our input `X` through the dropout operation, with probabilities 0, 0.5, and 1, respectively.

```
X = torch.arange(16, dtype = torch.float32).reshape((2, 8))
print('dropout_p = 0:', dropout_layer(X, 0))
print('dropout_p = 0.5:', dropout_layer(X, 0.5))
print('dropout_p = 1:', dropout_layer(X, 1))
```

```
dropout_p = 0: tensor([[ 0.,  1.,  2.,  3.,  4.,  5.,  6.,  7.],
        [ 8.,  9., 10., 11., 12., 13., 14., 15.]])
dropout_p = 0.5: tensor([[ 0.,  2.,  4.,  0.,  8., 10., 12.,  0.],
        [16., 18., 20., 22.,  0.,  0.,  0., 30.]])
dropout_p = 1: tensor([[0., 0., 0., 0., 0., 0., 0., 0.],
        [0., 0., 0., 0., 0., 0., 0., 0.]])
```

### Defining the Model

The model below applies dropout to the output of each hidden layer (following the activation function). We can set dropout probabilities for each layer separately. A common choice is to set a lower dropout probability closer to the input layer. We ensure that dropout is only active during training.

```
class DropoutMLPScratch(d2l.Classifier):
    def __init__(self, num_outputs, num_hiddens_1, num_hiddens_2,
                 dropout_1, dropout_2, lr):
        super().__init__()
        self.save_hyperparameters()
        self.lin1 = nn.LazyLinear(num_hiddens_1)
        self.lin2 = nn.LazyLinear(num_hiddens_2)
        self.lin3 = nn.LazyLinear(num_outputs)
        self.relu = nn.ReLU()

    def forward(self, X):
        H1 = self.relu(self.lin1(X.reshape((X.shape[0], -1))))
        if self.training:
            H1 = dropout_layer(H1, self.dropout_1)
        H2 = self.relu(self.lin2(H1))
        if self.training:
            H2 = dropout_layer(H2, self.dropout_2)
        return self.lin3(H2)
```

### Training

The following is similar to the training of MLPs described previously.

```
hparams = {'num_outputs':10, 'num_hiddens_1':256, 'num_hiddens_2':256,
           'dropout_1':0.5, 'dropout_2':0.5, 'lr':0.1}
model = DropoutMLPScratch(**hparams)
data = d2l.FashionMNIST(batch_size=256)
```

```
trainer = d2l.Trainer(max_epochs=10)
trainer.fit(model, data)
```

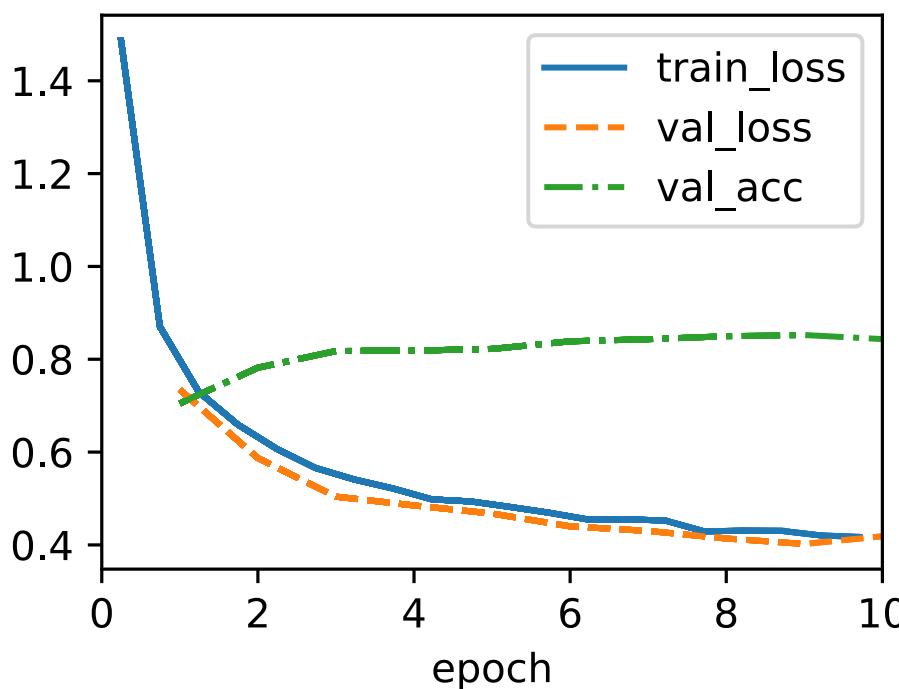

## 5.6.3 Concise Implementation

With high-level APIs, all we need to do is add a `Dropout` layer after each fully connected layer, passing in the dropout probability as the only argument to its constructor. During training, the `Dropout` layer will randomly drop out outputs of the previous layer (or equivalently, the inputs to the subsequent layer) according to the specified dropout probability. When not in training mode, the `Dropout` layer simply passes the data through during testing.

```
class DropoutMLP(d2l.Classifier):
    def __init__(self, num_outputs, num_hiddens_1, num_hiddens_2,
                 dropout_1, dropout_2, lr):
        super().__init__()
        self.save_hyperparameters()
        self.net = nn.Sequential(
            nn.Flatten(), nn.LazyLinear(num_hiddens_1), nn.ReLU(),
            nn.Dropout(dropout_1), nn.LazyLinear(num_hiddens_2), nn.ReLU(),
            nn.Dropout(dropout_2), nn.LazyLinear(num_outputs))
```

Next, we train the model.

```
model = DropoutMLP(**hparams)
trainer.fit(model, data)
```

## 5.6.4 Summary

Beyond controlling the number of dimensions and the size of the weight vector, dropout is yet another tool for avoiding overfitting. Often tools are used jointly. Note that dropout is used only during training: it replaces an activation $h$ with a random variable with expected value $h$.

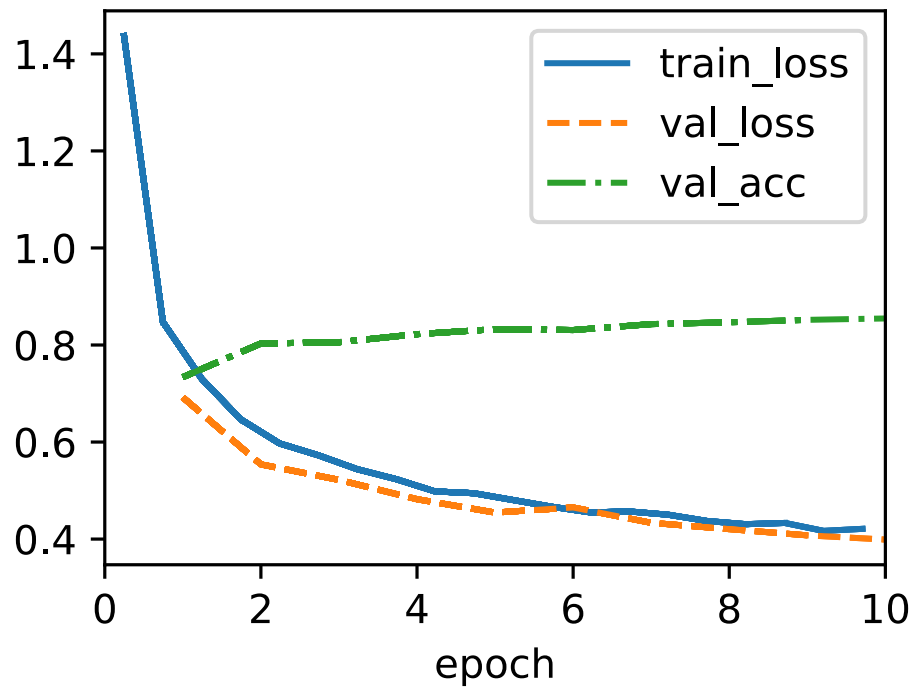

## 5.6.5 Exercises

1. What happens if you change the dropout probabilities for the first and second layers? In particular, what happens if you switch the ones for both layers? Design an experiment to answer these questions, describe your results quantitatively, and summarize the qualitative takeaways.

2. Increase the number of epochs and compare the results obtained when using dropout with those when not using it.

3. What is the variance of the activations in each hidden layer when dropout is and is not applied? Draw a plot to show how this quantity evolves over time for both models.

4. Why is dropout not typically used at test time?

5. Using the model in this section as an example, compare the effects of using dropout and weight decay. What happens when dropout and weight decay are used at the same time? Are the results additive? Are there diminished returns (or worse)? Do they cancel each other out?

6. What happens if we apply dropout to the individual weights of the weight matrix rather than the activations?

7. Invent another technique for injecting random noise at each layer that is different from the standard dropout technique. Can you develop a method that outperforms dropout on the Fashion-MNIST dataset (for a fixed architecture)?

Discussions[107].

## 5.7 Predicting House Prices on Kaggle

Now that we have introduced some basic tools for building and training deep networks and regularizing them with techniques including weight decay and dropout, we are ready to put all this knowledge into practice by participating in a Kaggle competition. The house price prediction competition is a great place to start. The data is fairly generic and do not exhibit exotic structure that might require specialized models (as audio or video might). This dataset, collected by De Cock (2011), covers house prices in Ames, Iowa from the period 2006–2010. It is considerably larger than the famous Boston housing dataset[108] of Harrison and Rubinfeld (1978), boasting both more examples and more features.

In this section, we will walk you through details of data preprocessing, model design, and hyperparameter selection. We hope that through a hands-on approach, you will gain some intuitions that will guide you in your career as a data scientist.

```
%matplotlib inline
import pandas as pd
import torch
from torch import nn
from d2l import torch as d2l
```

### 5.7.1 Downloading Data

Throughout the book, we will train and test models on various downloaded datasets. Here, we implement two utility functions for downloading and extracting zip or tar files. Again, we skip implementation details of such utility functions.

```
def download(url, folder, sha1_hash=None):
    """Download a file to folder and return the local filepath."""

def extract(filename, folder):
    """Extract a zip/tar file into folder."""
```

### 5.7.2 Kaggle

Kaggle[109] is a popular platform that hosts machine learning competitions. Each competition centers on a dataset and many are sponsored by stakeholders who offer prizes to the winning solutions. The platform helps users to interact via forums and shared code, fostering both collaboration and competition. While leaderboard chasing often spirals out of control, with researchers focusing myopically on preprocessing steps rather than asking fundamental questions, there is also tremendous value in the objectivity of a platform that facilitates direct quantitative comparisons among competing approaches as well as code sharing so that everyone can learn what did and did not work. If you want to participate in a Kaggle competition, you will first need to register for an account (see Fig. 5.7.1).

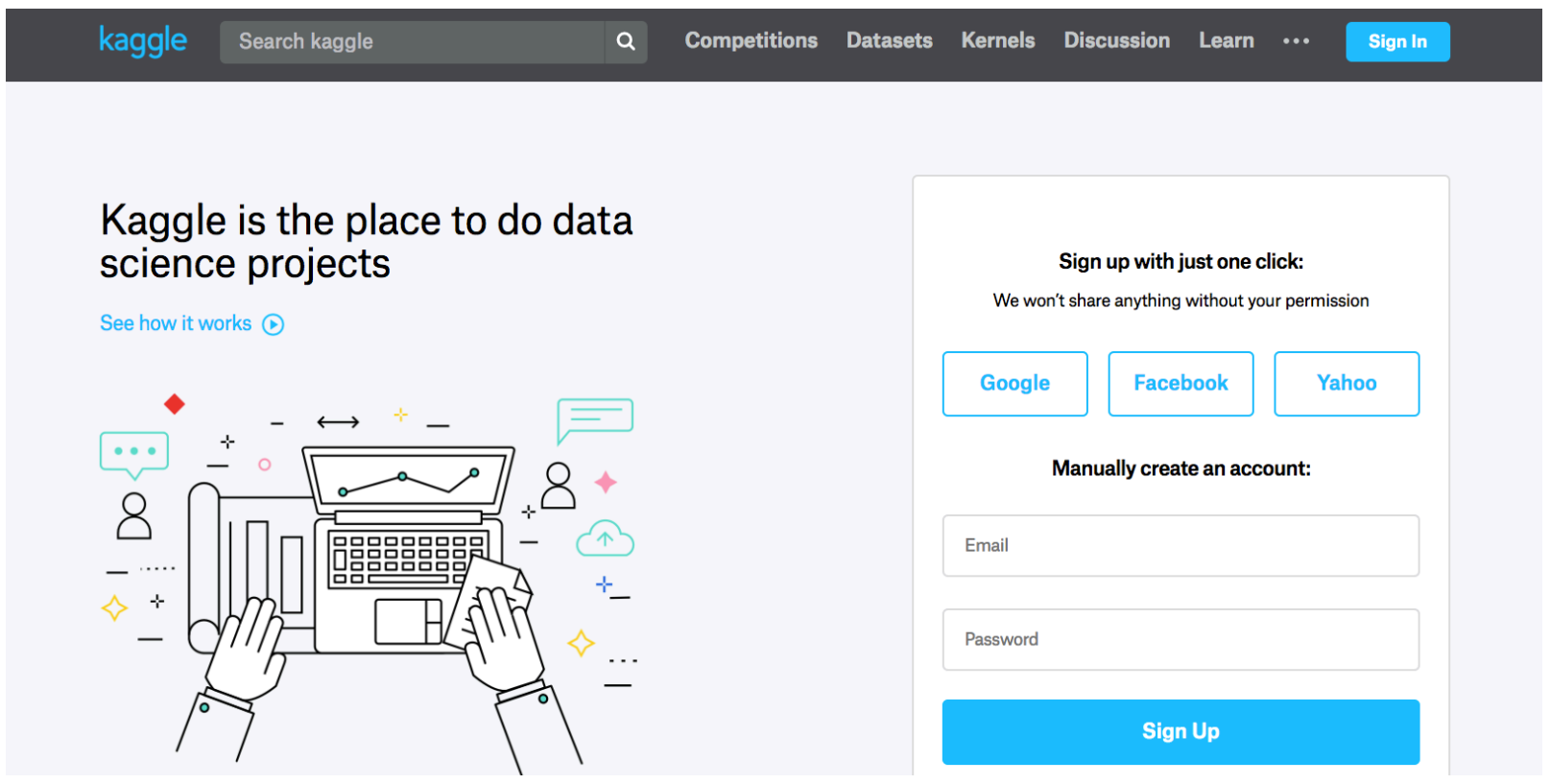

Fig. 5.7.1 The Kaggle website.

On the house price prediction competition page, as illustrated in Fig. 5.7.2, you can find the dataset (under the "Data" tab), submit predictions, and see your ranking, The URL is right here:

https://www.kaggle.com/c/house-prices-advanced-regression-techniques

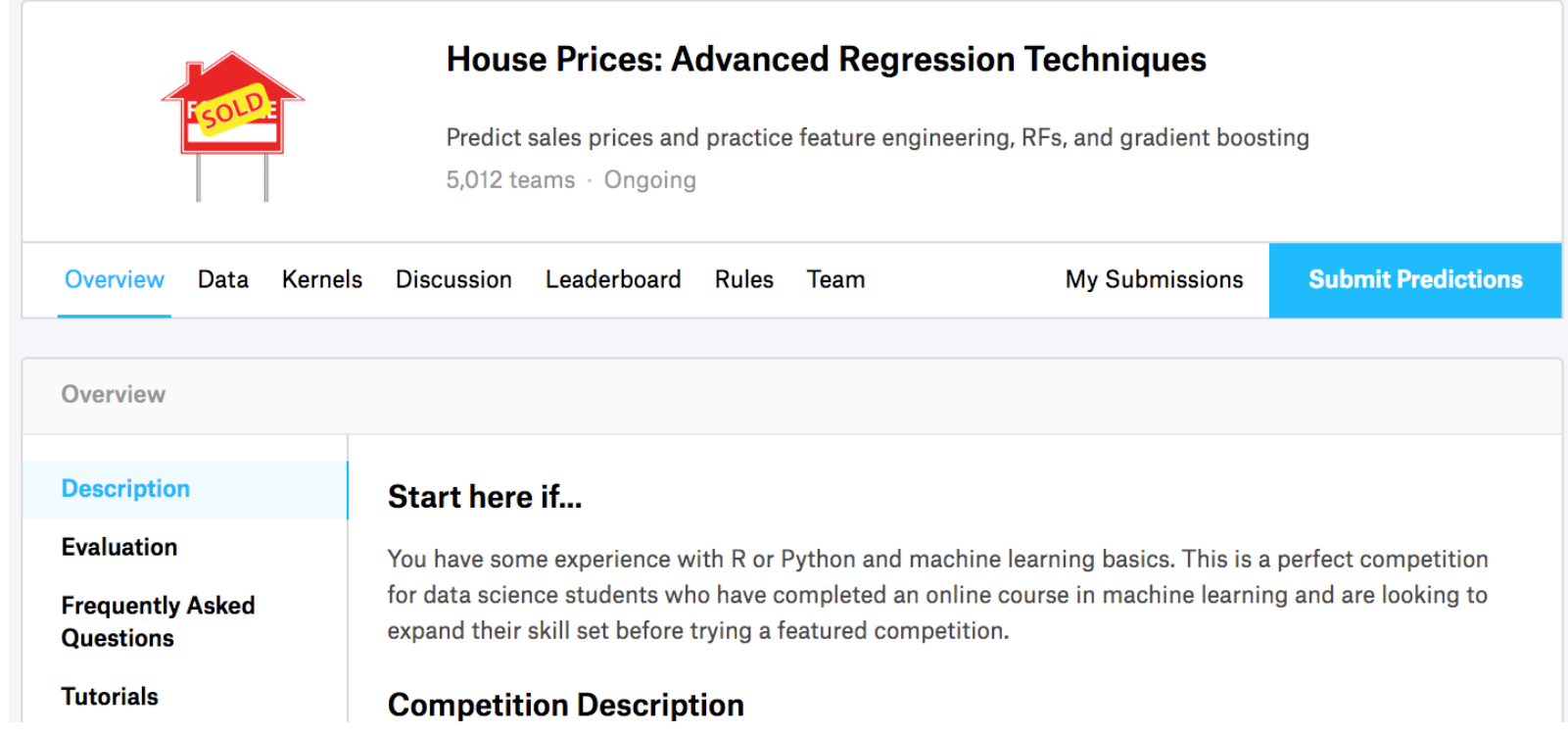

Fig. 5.7.2 The house price prediction competition page.

## 5.7.3 Accessing and Reading the Dataset

Note that the competition data is separated into training and test sets. Each record includes the property value of the house and attributes such as street type, year of construction, roof type, basement condition, etc. The features consist of various data types. For example, the year of construction is represented by an integer, the roof type by discrete categorical assignments, and other features by floating point numbers. And here is where reality complicates things:

for some examples, some data is altogether missing with the missing value marked simply as "na". The price of each house is included for the training set only (it is a competition after all). We will want to partition the training set to create a validation set, but we only get to evaluate our models on the official test set after uploading predictions to Kaggle. The "Data" tab on the competition tab in Fig. 5.7.2 has links for downloading the data.

To get started, we will read in and process the data using `pandas`, which we introduced in Section 2.2. For convenience, we can download and cache the Kaggle housing dataset. If a file corresponding to this dataset already exists in the cache directory and its SHA-1 matches `sha1_hash`, our code will use the cached file to avoid clogging up your Internet with redundant downloads.

```
class KaggleHouse(d2l.DataModule):
    def __init__(self, batch_size, train=None, val=None):
        super().__init__()
        self.save_hyperparameters()
        if self.train is None:
            self.raw_train = pd.read_csv(d2l.download(
                d2l.DATA_URL + 'kaggle_house_pred_train.csv', self.root,
                sha1_hash='585e9cc93e70b39160e7921475f9bcd7d31219ce'))
            self.raw_val = pd.read_csv(d2l.download(
                d2l.DATA_URL + 'kaggle_house_pred_test.csv', self.root,
                sha1_hash='fa19780a7b011d9b009e8bff8e99922a8ee2eb90'))
```

The training dataset includes 1460 examples, 80 features, and one label, while the validation data contains 1459 examples and 80 features.

```
data = KaggleHouse(batch_size=64)
print(data.raw_train.shape)
print(data.raw_val.shape)
```

```
(1460, 81)
(1459, 80)
```

## 5.7.4 Data Preprocessing

Let's take a look at the first four and final two features as well as the label (SalePrice) from the first four examples.

```
print(data.raw_train.iloc[:4, [0, 1, 2, 3, -3, -2, -1]])
```

```
   Id  MSSubClass MSZoning  LotFrontage SaleType SaleCondition  SalePrice
0   1          60       RL         65.0       WD        Normal     208500
1   2          20       RL         80.0       WD        Normal     181500
2   3          60       RL         68.0       WD        Normal     223500
3   4          70       RL         60.0       WD       Abnorml     140000
```

We can see that in each example, the first feature is the identifier. This helps the model determine each training example. While this is convenient, it does not carry any information for prediction purposes. Hence, we will remove it from the dataset before feeding the data into the model. Furthermore, given a wide variety of data types, we will need to preprocess the data before we can start modeling.

Let's start with the numerical features. First, we apply a heuristic, replacing all missing values by the corresponding feature's mean. Then, to put all features on a common scale, we *standardize* the data by rescaling features to zero mean and unit variance:

$$x \leftarrow \frac{x - \mu}{\sigma}, \tag{5.7.1}$$

where $\mu$ and $\sigma$ denote mean and standard deviation, respectively. To verify that this indeed transforms our feature (variable) such that it has zero mean and unit variance, note that $E[\frac{x-\mu}{\sigma}] = \frac{\mu-\mu}{\sigma} = 0$ and that $E[(x-\mu)^2] = (\sigma^2 + \mu^2) - 2\mu^2 + \mu^2 = \sigma^2$. Intuitively, we standardize the data for two reasons. First, it proves convenient for optimization. Second, because we do not know *a priori* which features will be relevant, we do not want to penalize coefficients assigned to one feature more than any other.

Next we deal with discrete values. These include features such as "MSZoning". We replace them by a one-hot encoding in the same way that we earlier transformed multiclass labels into vectors (see Section 4.1.1). For instance, "MSZoning" assumes the values "RL" and "RM". Dropping the "MSZoning" feature, two new indicator features "MSZoning_RL" and "MSZoning_RM" are created with values being either 0 or 1. According to one-hot encoding, if the original value of "MSZoning" is "RL", then "MSZoning_RL" is 1 and "MSZoning_RM" is 0. The `pandas` package does this automatically for us.

```python
@d2l.add_to_class(KaggleHouse)
def preprocess(self):
    # Remove the ID and label columns
    label = 'SalePrice'
    features = pd.concat(
        (self.raw_train.drop(columns=['Id', label]),
         self.raw_val.drop(columns=['Id'])))
    # Standardize numerical columns
    numeric_features = features.dtypes[features.dtypes!='object'].index
    features[numeric_features] = features[numeric_features].apply(
        lambda x: (x - x.mean()) / (x.std()))
    # Replace NAN numerical features by 0
    features[numeric_features] = features[numeric_features].fillna(0)
    # Replace discrete features by one-hot encoding
    features = pd.get_dummies(features, dummy_na=True)
    # Save preprocessed features
    self.train = features[:self.raw_train.shape[0]].copy()
    self.train[label] = self.raw_train[label]
    self.val = features[self.raw_train.shape[0]:].copy()
```

You can see that this conversion increases the number of features from 79 to 331 (excluding ID and label columns).

```
data.preprocess()
data.train.shape
```

```
(1460, 331)
```

### 5.7.5 Error Measure

To get started we will train a linear model with squared loss. Not surprisingly, our linear model will not lead to a competition-winning submission but it does provide a sanity check to see whether there is meaningful information in the data. If we cannot do better than random guessing here, then there might be a good chance that we have a data processing bug. And if things work, the linear model will serve as a baseline giving us some intuition about how close the simple model gets to the best reported models, giving us a sense of how much gain we should expect from fancier models.

With house prices, as with stock prices, we care about relative quantities more than absolute quantities. Thus we tend to care more about the relative error $\frac{y-\hat{y}}{y}$ than about the absolute error $y - \hat{y}$. For instance, if our prediction is off by $100,000 when estimating the price of a house in rural Ohio, where the value of a typical house is $125,000, then we are probably doing a horrible job. On the other hand, if we err by this amount in Los Altos Hills, California, this might represent a stunningly accurate prediction (there, the median house price exceeds $4 million).

One way to address this problem is to measure the discrepancy in the logarithm of the price estimates. In fact, this is also the official error measure used by the competition to evaluate the quality of submissions. After all, a small value $\delta$ for $|\log y - \log \hat{y}| \leq \delta$ translates into $e^{-\delta} \leq \frac{\hat{y}}{y} \leq e^{\delta}$. This leads to the following root-mean-squared-error between the logarithm of the predicted price and the logarithm of the label price:

$$\sqrt{\frac{1}{n}\sum_{i=1}^n\left(\log y_i -\log \hat{y}_i\right)^2}. \tag{5.7.2}$$

```
@d2l.add_to_class(KaggleHouse)
def get_dataloader(self, train):
    label = 'SalePrice'
    data = self.train if train else self.val
    if label not in data: return
    get_tensor = lambda x: torch.tensor(x.values.astype(float),
                                      dtype=torch.float32)
    # Logarithm of prices
    tensors = (get_tensor(data.drop(columns=[label])),  # X
               torch.log(get_tensor(data[label])).reshape((-1, 1)))  # Y
    return self.get_tensorloader(tensors, train)
```

### 5.7.6 $K$-Fold Cross-Validation

You might recall that we introduced cross-validation in Section 3.6.3, where we discussed how to deal with model selection. We will put this to good use to select the model design and to adjust the hyperparameters. We first need a function that returns the $i^{\mathrm{th}}$ fold of the data in a $K$-fold cross-validation procedure. It proceeds by slicing out the $i^{\mathrm{th}}$ segment as validation data and returning the rest as training data. Note that this is not the most efficient way of handling data and we would definitely do something much smarter if our dataset was considerably larger. But this added complexity might obfuscate our code unnecessarily so we can safely omit it here owing to the simplicity of our problem.

```
def k_fold_data(data, k):
    rets = []
    fold_size = data.train.shape[0] // k
    for j in range(k):
        idx = range(j * fold_size, (j+1) * fold_size)
        rets.append(KaggleHouse(data.batch_size, data.train.drop(index=idx),
                                data.train.loc[idx]))
    return rets
```

The average validation error is returned when we train $K$ times in the $K$-fold cross-validation.

```
def k_fold(trainer, data, k, lr):
    val_loss, models = [], []
    for i, data_fold in enumerate(k_fold_data(data, k)):
        model = d2l.LinearRegression(lr)
        model.board.yscale='log'
        if i != 0: model.board.display = False
        trainer.fit(model, data_fold)
        val_loss.append(float(model.board.data['val_loss'][-1].y))
        models.append(model)
    print(f'average validation log mse = {sum(val_loss)/len(val_loss)}')
    return models
```

### 5.7.7 Model Selection

In this example, we pick an untuned set of hyperparameters and leave it up to the reader to improve the model. Finding a good choice can take time, depending on how many variables one optimizes over. With a large enough dataset, and the normal sorts of hyperparameters, $K$-fold cross-validation tends to be reasonably resilient against multiple testing. However, if we try an unreasonably large number of options we might find that our validation performance is no longer representative of the true error.

```
trainer = d2l.Trainer(max_epochs=10)
models = k_fold(trainer, data, k=5, lr=0.01)
```

```
average validation log mse = 0.18610747516155243
```

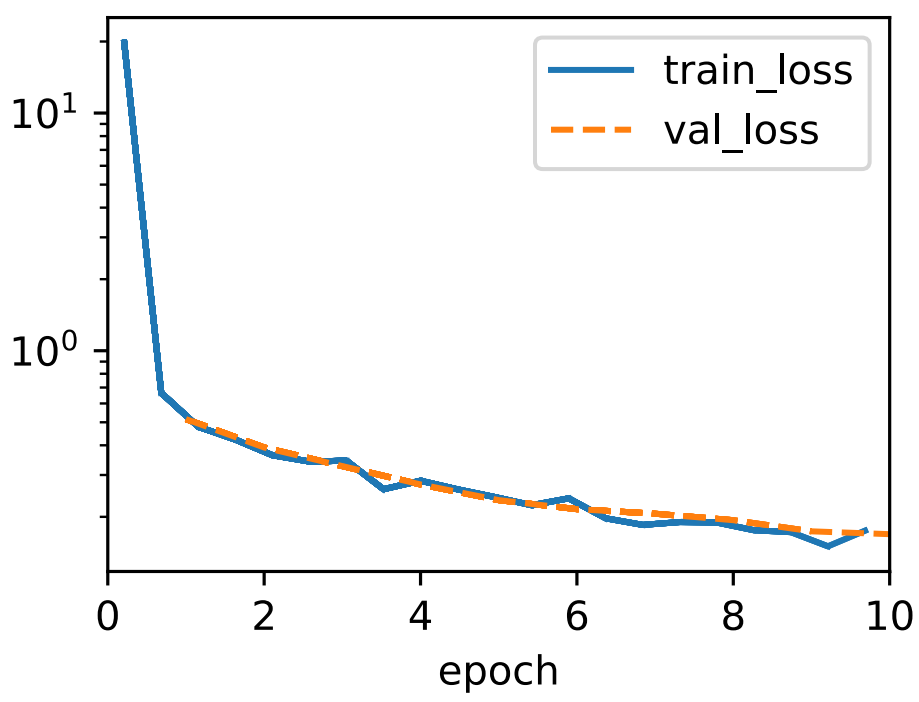

Notice that sometimes the number of training errors for a set of hyperparameters can be very low, even as the number of errors on $K$-fold cross-validation grows considerably higher. This indicates that we are overfitting. Throughout training you will want to monitor both numbers. Less overfitting might indicate that our data can support a more powerful model. Massive overfitting might suggest that we can gain by incorporating regularization techniques.

### 5.7.8 Submitting Predictions on Kaggle

Now that we know what a good choice of hyperparameters should be, we might calculate the average predictions on the test set by all the $K$ models. Saving the predictions in a csv file will simplify uploading the results to Kaggle. The following code will generate a file called `submission.csv`.

```
preds = [model(torch.tensor(data.val.values.astype(float), dtype=torch.
↪float32))
         for model in models]
# Taking exponentiation of predictions in the logarithm scale
ensemble_preds = torch.exp(torch.cat(preds, 1)).mean(1)
submission = pd.DataFrame({'Id':data.raw_val.Id,
                           'SalePrice':ensemble_preds.detach().numpy()})
submission.to_csv('submission.csv', index=False)
```

Next, as demonstrated in Fig. 5.7.3, we can submit our predictions on Kaggle and see how they compare with the actual house prices (labels) on the test set. The steps are quite simple:

- Log in to the Kaggle website and visit the house price prediction competition page.
- Click the “Submit Predictions” or “Late Submission” button.
- Click the “Upload Submission File” button in the dashed box at the bottom of the page and select the prediction file you wish to upload.

- Click the "Make Submission" button at the bottom of the page to view your results.

Fig. 5.7.3 Submitting data to Kaggle

## 5.7.9 Summary and Discussion

Real data often contains a mix of different data types and needs to be preprocessed. Rescaling real-valued data to zero mean and unit variance is a good default. So is replacing missing values with their mean. Furthermore, transforming categorical features into indicator features allows us to treat them like one-hot vectors. When we tend to care more about the relative error than about the absolute error, we can measure the discrepancy in the logarithm of the prediction. To select the model and adjust the hyperparameters, we can use $K$-fold cross-validation .

## 5.7.10 Exercises

1. Submit your predictions for this section to Kaggle. How good are they?
2. Is it always a good idea to replace missing values by a mean? Hint: can you construct a situation where the values are not missing at random?
3. Improve the score by tuning the hyperparameters through $K$-fold cross-validation.
4. Improve the score by improving the model (e.g., layers, weight decay, and dropout).
5. What happens if we do not standardize the continuous numerical features as we have done in this section?

Discussions[110].

# 6 Builders' Guide

Alongside giant datasets and powerful hardware, great software tools have played an indispensable role in the rapid progress of deep learning. Starting with the pathbreaking Theano library released in 2007, flexible open-source tools have enabled researchers to rapidly prototype models, avoiding repetitive work when recycling standard components while still maintaining the ability to make low-level modifications. Over time, deep learning's libraries have evolved to offer increasingly coarse abstractions. Just as semiconductor designers went from specifying transistors to logical circuits to writing code, neural networks researchers have moved from thinking about the behavior of individual artificial neurons to conceiving of networks in terms of whole layers, and now often design architectures with far coarser *blocks* in mind.

So far, we have introduced some basic machine learning concepts, ramping up to fully-functional deep learning models. In the last chapter, we implemented each component of an MLP from scratch and even showed how to leverage high-level APIs to roll out the same models effortlessly. To get you that far that fast, we *called upon* the libraries, but skipped over more advanced details about *how they work*. In this chapter, we will peel back the curtain, digging deeper into the key components of deep learning computation, namely model construction, parameter access and initialization, designing custom layers and blocks, reading and writing models to disk, and leveraging GPUs to achieve dramatic speedups. These insights will move you from *end user* to *power user*, giving you the tools needed to reap the benefits of a mature deep learning library while retaining the flexibility to implement more complex models, including those you invent yourself! While this chapter does not introduce any new models or datasets, the advanced modeling chapters that follow rely heavily on these techniques.

## 6.1 Layers and Modules

When we first introduced neural networks, we focused on linear models with a single output. Here, the entire model consists of just a single neuron. Note that a single neuron (i) takes some set of inputs; (ii) generates a corresponding scalar output; and (iii) has a set of associated parameters that can be updated to optimize some objective function of interest. Then, once we started thinking about networks with multiple outputs, we leveraged vectorized arithmetic to characterize an entire layer of neurons. Just like individual neurons, layers (i) take a set of inputs, (ii) generate corresponding outputs, and (iii) are described by a set of tunable parameters. When we worked through softmax regression, a single layer was itself the model. However, even when we subsequently introduced MLPs, we could still think of the model as retaining this same basic structure.

Interestingly, for MLPs, both the entire model and its constituent layers share this structure. The entire model takes in raw inputs (the features), generates outputs (the predictions), and possesses parameters (the combined parameters from all constituent layers). Likewise, each individual layer ingests inputs (supplied by the previous layer) generates outputs (the inputs to the subsequent layer), and possesses a set of tunable parameters that are updated according to the signal that flows backwards from the subsequent layer.

While you might think that neurons, layers, and models give us enough abstractions to go about our business, it turns out that we often find it convenient to speak about components that are larger than an individual layer but smaller than the entire model. For example, the ResNet-152 architecture, which is wildly popular in computer vision, possesses hundreds of layers. These layers consist of repeating patterns of *groups of layers*. Implementing such a network one layer at a time can grow tedious. This concern is not just hypothetical—such design patterns are common in practice. The ResNet architecture mentioned above won the 2015 ImageNet and COCO computer vision competitions for both recognition and detection (He *et al.*, 2016) and remains a go-to architecture for many vision tasks. Similar architectures in which layers are arranged in various repeating patterns are now ubiquitous in other domains, including natural language processing and speech.

To implement these complex networks, we introduce the concept of a neural network *module*. A module could describe a single layer, a component consisting of multiple layers, or the entire model itself! One benefit of working with the module abstraction is that they can be combined into larger artifacts, often recursively. This is illustrated in Fig. 6.1.1. By defining code to generate modules of arbitrary complexity on demand, we can write surprisingly compact code and still implement complex neural networks.

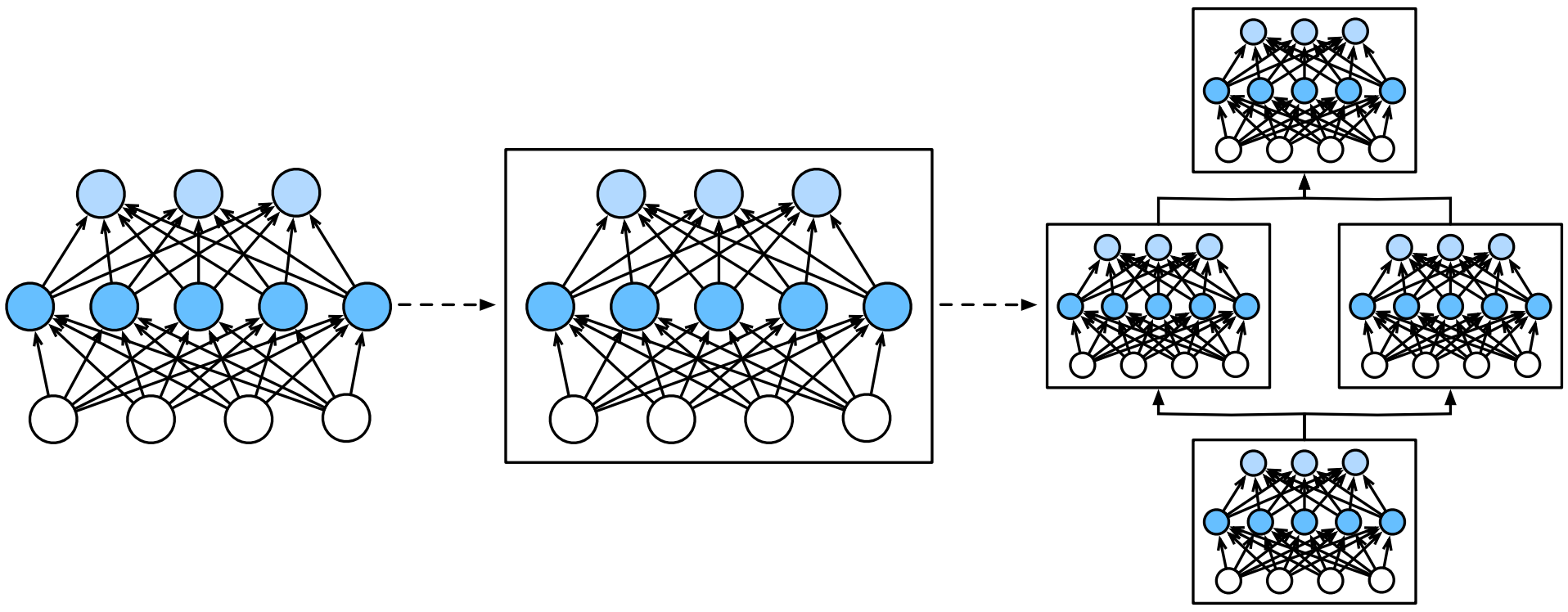

Fig. 6.1.1 Multiple layers are combined into modules, forming repeating patterns of larger models.

From a programming standpoint, a module is represented by a *class*. Any subclass of it must define a forward propagation method that transforms its input into output and must store any necessary parameters. Note that some modules do not require any parameters at all. Finally a module must possess a backpropagation method, for purposes of calculating gradients. Fortunately, due to some behind-the-scenes magic supplied by the auto differentiation (introduced in Section 2.5) when defining our own module, we only need to worry about parameters and the forward propagation method.

```
import torch
from torch import nn
from torch.nn import functional as F
```

To begin, we revisit the code that we used to implement MLPs (Section 5.1). The following code generates a network with one fully connected hidden layer with 256 units and ReLU activation, followed by a fully connected output layer with ten units (no activation function).

```
net = nn.Sequential(nn.LazyLinear(256), nn.ReLU(), nn.LazyLinear(10))

X = torch.rand(2, 20)
net(X).shape
```

```
torch.Size([2, 10])
```

In this example, we constructed our model by instantiating an `nn.Sequential`, with layers in the order that they should be executed passed as arguments. In short, `nn.Sequential` defines a special kind of `Module`, the class that presents a module in PyTorch. It maintains an ordered list of constituent `Module`s. Note that each of the two fully connected layers is an instance of the `Linear` class which is itself a subclass of `Module`. The forward propagation (`forward`) method is also remarkably simple: it chains each module in the list together, passing the output of each as input to the next. Note that until now, we have been invoking our models via the construction `net(X)` to obtain their outputs. This is actually just shorthand for `net.__call__(X)`.

### 6.1.1 A Custom Module

Perhaps the easiest way to develop intuition about how a module works is to implement one ourselves. Before we do that, we briefly summarize the basic functionality that each module must provide:

1. Ingest input data as arguments to its forward propagation method.
2. Generate an output by having the forward propagation method return a value. Note that the output may have a different shape from the input. For example, the first fully connected layer in our model above ingests an input of arbitrary dimension but returns an output of dimension 256.

3. Calculate the gradient of its output with respect to its input, which can be accessed via its backpropagation method. Typically this happens automatically.

4. Store and provide access to those parameters necessary for executing the forward propagation computation.

5. Initialize model parameters as needed.

In the following snippet, we code up a module from scratch corresponding to an MLP with one hidden layer with 256 hidden units, and a 10-dimensional output layer. Note that the `MLP` class below inherits the class that represents a module. We will heavily rely on the parent class's methods, supplying only our own constructor (the `__init__` method in Python) and the forward propagation method.

```
class MLP(nn.Module):
    def __init__(self):
        # Call the constructor of the parent class nn.Module to perform
        # the necessary initialization
        super().__init__()
        self.hidden = nn.LazyLinear(256)
        self.out = nn.LazyLinear(10)

    # Define the forward propagation of the model, that is, how to return the
    # required model output based on the input X
    def forward(self, X):
        return self.out(F.relu(self.hidden(X)))
```

Let's first focus on the forward propagation method. Note that it takes `X` as input, calculates the hidden representation with the activation function applied, and outputs its logits. In this `MLP` implementation, both layers are instance variables. To see why this is reasonable, imagine instantiating two MLPs, `net1` and `net2`, and training them on different data. Naturally, we would expect them to represent two different learned models.

We instantiate the MLP's layers in the constructor and subsequently invoke these layers on each call to the forward propagation method. Note a few key details. First, our customized `__init__` method invokes the parent class's `__init__` method via `super().__init__()` sparing us the pain of restating boilerplate code applicable to most modules. We then instantiate our two fully connected layers, assigning them to `self.hidden` and `self.out`. Note that unless we implement a new layer, we need not worry about the backpropagation method or parameter initialization. The system will generate these methods automatically. Let's try this out.

```
net = MLP()
net(X).shape
```

```
torch.Size([2, 10])
```

A key virtue of the module abstraction is its versatility. We can subclass a module to create layers (such as the fully connected layer class), entire models (such as the `MLP` class above), or various components of intermediate complexity. We exploit this versatility throughout the coming chapters, such as when addressing convolutional neural networks.

### 6.1.2 The Sequential Module

We can now take a closer look at how the `Sequential` class works. Recall that `Sequential` was designed to daisy-chain other modules together. To build our own simplified `MySequential`, we just need to define two key methods:

1. A method for appending modules one by one to a list.
2. A forward propagation method for passing an input through the chain of modules, in the same order as they were appended.

The following `MySequential` class delivers the same functionality of the default `Sequential` class.

```
class MySequential(nn.Module):
    def __init__(self, *args):
        super().__init__()
        for idx, module in enumerate(args):
            self.add_module(str(idx), module)

    def forward(self, X):
        for module in self.children():
            X = module(X)
        return X
```

In the `__init__` method, we add every module by calling the `add_modules` method. These modules can be accessed by the `children` method at a later date. In this way the system knows the added modules, and it will properly initialize each module's parameters.

When our `MySequential`'s forward propagation method is invoked, each added module is executed in the order in which they were added. We can now reimplement an MLP using our `MySequential` class.

```
net = MySequential(nn.LazyLinear(256), nn.ReLU(), nn.LazyLinear(10))
net(X).shape
```

```
torch.Size([2, 10])
```

Note that this use of `MySequential` is identical to the code we previously wrote for the `Sequential` class (as described in Section 5.1).

### 6.1.3 Executing Code in the Forward Propagation Method

The `Sequential` class makes model construction easy, allowing us to assemble new architectures without having to define our own class. However, not all architectures are simple daisy chains. When greater flexibility is required, we will want to define our own blocks. For example, we might want to execute Python's control flow within the forward propagation method. Moreover, we might want to perform arbitrary mathematical operations, not simply relying on predefined neural network layers.

You may have noticed that until now, all of the operations in our networks have acted upon our network's activations and its parameters. Sometimes, however, we might want to incorporate terms that are neither the result of previous layers nor updatable parameters. We call these *constant parameters*. Say for example that we want a layer that calculates the function $f(\mathbf{x}, \mathbf{w}) = c \cdot \mathbf{w}^\top \mathbf{x}$, where $\mathbf{x}$ is the input, $\mathbf{w}$ is our parameter, and $c$ is some specified constant that is not updated during optimization. So we implement a `FixedHiddenMLP` class as follows.

```
class FixedHiddenMLP(nn.Module):
    def __init__(self):
        super().__init__()
        # Random weight parameters that will not compute gradients and
        # therefore keep constant during training
        self.rand_weight = torch.rand((20, 20))
        self.linear = nn.LazyLinear(20)

    def forward(self, X):
        X = self.linear(X)
        X = F.relu(X @ self.rand_weight + 1)
        # Reuse the fully connected layer. This is equivalent to sharing
        # parameters with two fully connected layers
        X = self.linear(X)
        # Control flow
        while X.abs().sum() > 1:
            X /= 2
        return X.sum()
```

In this model, we implement a hidden layer whose weights (`self.rand_weight`) are initialized randomly at instantiation and are thereafter constant. This weight is not a model parameter and thus it is never updated by backpropagation. The network then passes the output of this "fixed" layer through a fully connected layer.

Note that before returning the output, our model did something unusual. We ran a while-loop, testing on the condition its $\ell_1$ norm is larger than $1$, and dividing our output vector by $2$ until it satisfied the condition. Finally, we returned the sum of the entries in `X`. To our knowledge, no standard neural network performs this operation. Note that this particular operation may not be useful in any real-world task. Our point is only to show you how to integrate arbitrary code into the flow of your neural network computations.

```
net = FixedHiddenMLP()
net(X)
```

```
tensor(-0.1635, grad_fn=<SumBackward0>)
```

We can mix and match various ways of assembling modules together. In the following example, we nest modules in some creative ways.

```
class NestMLP(nn.Module):
    def __init__(self):
        super().__init__()
        self.net = nn.Sequential(nn.LazyLinear(64), nn.ReLU(),
                                 nn.LazyLinear(32), nn.ReLU())
        self.linear = nn.LazyLinear(16)

    def forward(self, X):
        return self.linear(self.net(X))

chimera = nn.Sequential(NestMLP(), nn.LazyLinear(20), FixedHiddenMLP())
chimera(X)
```

```
tensor(0.2447, grad_fn=<SumBackward0>)
```

### 6.1.4 Summary

Individual layers can be modules. Many layers can comprise a module. Many modules can comprise a module.

A module can contain code. Modules take care of lots of housekeeping, including parameter initialization and backpropagation. Sequential concatenations of layers and modules are handled by the `Sequential` module.

### 6.1.5 Exercises

1. What kinds of problems will occur if you change `MySequential` to store modules in a Python list?
2. Implement a module that takes two modules as an argument, say `net1` and `net2` and returns the concatenated output of both networks in the forward propagation. This is also called a *parallel module*.
3. Assume that you want to concatenate multiple instances of the same network. Implement a factory function that generates multiple instances of the same module and build a larger network from it.

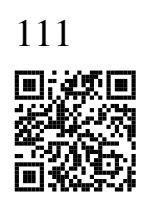

Discussions[111].

## 6.2 Parameter Management

Once we have chosen an architecture and set our hyperparameters, we proceed to the training loop, where our goal is to find parameter values that minimize our loss function. After training, we will need these parameters in order to make future predictions. Additionally, we will sometimes wish to extract the parameters perhaps to reuse them in some other context, to save our model to disk so that it may be executed in other software, or for examination in the hope of gaining scientific understanding.

Most of the time, we will be able to ignore the nitty-gritty details of how parameters are declared and manipulated, relying on deep learning frameworks to do the heavy lifting. However, when we move away from stacked architectures with standard layers, we will sometimes need to get into the weeds of declaring and manipulating parameters. In this section, we cover the following:

- Accessing parameters for debugging, diagnostics, and visualizations.
- Sharing parameters across different model components.

```
import torch
from torch import nn
```

We start by focusing on an MLP with one hidden layer.

```
net = nn.Sequential(nn.LazyLinear(8),
                    nn.ReLU(),
                    nn.LazyLinear(1))

X = torch.rand(size=(2, 4))
net(X).shape
```

```
torch.Size([2, 1])
```

### 6.2.1 Parameter Access

Let's start with how to access parameters from the models that you already know.

When a model is defined via the `Sequential` class, we can first access any layer by indexing into the model as though it were a list. Each layer's parameters are conveniently located in its attribute.

We can inspect the parameters of the second fully connected layer as follows.

```
net[2].state_dict()
```

```
OrderedDict([('weight',
              tensor([[ 0.1842, -0.0048, -0.1618, -0.0667,  0.1058,  0.2521, ␣
↪0.3208, -0.1132]])),
             ('bias', tensor([0.1387]))])
```

We can see that this fully connected layer contains two parameters, corresponding to that layer's weights and biases, respectively.

### Targeted Parameters

Note that each parameter is represented as an instance of the parameter class. To do anything useful with the parameters, we first need to access the underlying numerical values. There are several ways to do this. Some are simpler while others are more general. The following code extracts the bias from the second neural network layer, which returns a parameter class instance, and further accesses that parameter's value.

```
type(net[2].bias), net[2].bias.data
```

```
(torch.nn.parameter.Parameter, tensor([0.1387]))
```

Parameters are complex objects, containing values, gradients, and additional information. That is why we need to request the value explicitly.

In addition to the value, each parameter also allows us to access the gradient. Because we have not invoked backpropagation for this network yet, it is in its initial state.

```
net[2].weight.grad == None
```

```
True
```

### All Parameters at Once

When we need to perform operations on all parameters, accessing them one-by-one can grow tedious. The situation can grow especially unwieldy when we work with more complex, e.g., nested, modules, since we would need to recurse through the entire tree to extract each sub-module's parameters. Below we demonstrate accessing the parameters of all layers.

```
[(name, param.shape) for name, param in net.named_parameters()]
```

```
[('0.weight', torch.Size([8, 4])),
 ('0.bias', torch.Size([8])),
 ('2.weight', torch.Size([1, 8])),
 ('2.bias', torch.Size([1]))]
```

## 6.2.2 Tied Parameters

Often, we want to share parameters across multiple layers. Let's see how to do this elegantly. In the following we allocate a fully connected layer and then use its parameters specifically to set those of another layer. Here we need to run the forward propagation `net(X)` before accessing the parameters.

```
# We need to give the shared layer a name so that we can refer to its
# parameters
shared = nn.LazyLinear(8)
net = nn.Sequential(nn.LazyLinear(8), nn.ReLU(),
                    shared, nn.ReLU(),
                    shared, nn.ReLU(),
                    nn.LazyLinear(1))

net(X)
# Check whether the parameters are the same
print(net[2].weight.data[0] == net[4].weight.data[0])
net[2].weight.data[0, 0] = 100
# Make sure that they are actually the same object rather than just having the
# same value
print(net[2].weight.data[0] == net[4].weight.data[0])
```

```
tensor([True, True, True, True, True, True, True, True])
tensor([True, True, True, True, True, True, True, True])
```

This example shows that the parameters of the second and third layer are tied. They are not just equal, they are represented by the same exact tensor. Thus, if we change one of the parameters, the other one changes, too.

You might wonder, when parameters are tied what happens to the gradients? Since the model parameters contain gradients, the gradients of the second hidden layer and the third hidden layer are added together during backpropagation.

## 6.2.3 Summary

We have several ways of accessing and tying model parameters.

## 6.2.4 Exercises

1. Use the NestMLP model defined in Section 6.1 and access the parameters of the various layers.
2. Construct an MLP containing a shared parameter layer and train it. During the training process, observe the model parameters and gradients of each layer.
3. Why is sharing parameters a good idea?

Discussions[112].

## 6.3 Parameter Initialization

Now that we know how to access the parameters, let's look at how to initialize them properly. We discussed the need for proper initialization in Section 5.4. The deep learning framework provides default random initializations to its layers. However, we often want to initialize our weights according to various other protocols. The framework provides most commonly used protocols, and also allows to create a custom initializer.

```
import torch
from torch import nn
```

By default, PyTorch initializes weight and bias matrices uniformly by drawing from a range that is computed according to the input and output dimension. PyTorch's nn.init module provides a variety of preset initialization methods.

```
net = nn.Sequential(nn.LazyLinear(8), nn.ReLU(), nn.LazyLinear(1))
X = torch.rand(size=(2, 4))
net(X).shape
```

```
torch.Size([2, 1])
```

### 6.3.1 Built-in Initialization

Let's begin by calling on built-in initializers. The code below initializes all weight parameters as Gaussian random variables with standard deviation 0.01, while bias parameters are cleared to zero.

```
def init_normal(module):
    if type(module) == nn.Linear:
        nn.init.normal_(module.weight, mean=0, std=0.01)
```

```
        nn.init.zeros_(module.bias)

net.apply(init_normal)
net[0].weight.data[0], net[0].bias.data[0]
```

```
(tensor([-0.0073, -0.0050, -0.0133, -0.0025]), tensor(0.))
```

We can also initialize all the parameters to a given constant value (say, 1).

```
def init_constant(module):
    if type(module) == nn.Linear:
        nn.init.constant_(module.weight, 1)
        nn.init.zeros_(module.bias)

net.apply(init_constant)
net[0].weight.data[0], net[0].bias.data[0]
```

```
(tensor([1., 1., 1., 1.]), tensor(0.))
```

We can also apply different initializers for certain blocks. For example, below we initialize the first layer with the Xavier initializer and initialize the second layer to a constant value of 42.

```
def init_xavier(module):
    if type(module) == nn.Linear:
        nn.init.xavier_uniform_(module.weight)

def init_42(module):
    if type(module) == nn.Linear:
        nn.init.constant_(module.weight, 42)

net[0].apply(init_xavier)
net[2].apply(init_42)
print(net[0].weight.data[0])
print(net[2].weight.data)
```

```
tensor([ 0.2163, -0.2129, -0.6275,  0.5273])
tensor([[42., 42., 42., 42., 42., 42., 42., 42.]])
```

### Custom Initialization

Sometimes, the initialization methods we need are not provided by the deep learning framework. In the example below, we define an initializer for any weight parameter $w$ using the following strange distribution:

$$w \sim \begin{cases} U(5,10) & \text{with probability } \frac{1}{4} \\ 0 & \text{with probability } \frac{1}{2} \\ U(-10,-5) & \text{with probability } \frac{1}{4} \end{cases} \tag{6.3.1}$$

Again, we implement a `my_init` function to apply to `net`.

```
def my_init(module):
    if type(module) == nn.Linear:
        print("Init", *[(name, param.shape)
                        for name, param in module.named_parameters()][0])
        nn.init.uniform_(module.weight, -10, 10)
        module.weight.data *= module.weight.data.abs() >= 5

net.apply(my_init)
net[0].weight[:2]
```

```
Init weight torch.Size([8, 4])
Init weight torch.Size([1, 8])
```

```
tensor([[ 7.0262, -9.7639,  7.1699, -7.3955],
        [-9.7688,  7.5430,  0.0000, -0.0000]], grad_fn=<SliceBackward0>)
```

Note that we always have the option of setting parameters directly.

```
net[0].weight.data[:] += 1
net[0].weight.data[0, 0] = 42
net[0].weight.data[0]
```

```
tensor([42.0000, -8.7639,  8.1699, -6.3955])
```

## 6.3.2 Summary

We can initialize parameters using built-in and custom initializers.

## 6.3.3 Exercises

Look up the online documentation for more built-in initializers.

Discussions[113].

## 6.4 Lazy Initialization

So far, it might seem that we got away with being sloppy in setting up our networks. Specifically, we did the following unintuitive things, which might not seem like they should work:

- We defined the network architectures without specifying the input dimensionality.
- We added layers without specifying the output dimension of the previous layer.
- We even "initialized" these parameters before providing enough information to determine how many parameters our models should contain.

You might be surprised that our code runs at all. After all, there is no way the deep learning framework could tell what the input dimensionality of a network would be. The trick here is that the framework *defers initialization*, waiting until the first time we pass data through the model, to infer the sizes of each layer on the fly.

Later on, when working with convolutional neural networks, this technique will become even more convenient since the input dimensionality (e.g., the resolution of an image) will affect the dimensionality of each subsequent layer. Hence the ability to set parameters without the need to know, at the time of writing the code, the value of the dimension can greatly simplify the task of specifying and subsequently modifying our models. Next, we go deeper into the mechanics of initialization.

```
import torch
from torch import nn
from d2l import torch as d2l
```

To begin, let's instantiate an MLP.

```
net = nn.Sequential(nn.LazyLinear(256), nn.ReLU(), nn.LazyLinear(10))
```

At this point, the network cannot possibly know the dimensions of the input layer's weights because the input dimension remains unknown.

Consequently the framework has not yet initialized any parameters. We confirm by attempting to access the parameters below.

```
net[0].weight
```

```
<UninitializedParameter>
```

Next let's pass data through the network to make the framework finally initialize parameters.

```
X = torch.rand(2, 20)
net(X)

net[0].weight.shape
```

```
torch.Size([256, 20])
```

As soon as we know the input dimensionality, 20, the framework can identify the shape of the first layer's weight matrix by plugging in the value of 20. Having recognized the first layer's shape, the framework proceeds to the second layer, and so on through the computational graph until all shapes are known. Note that in this case, only the first layer requires lazy initialization, but the framework initializes sequentially. Once all parameter shapes are known, the framework can finally initialize the parameters.

The following method passes in dummy inputs through the network for a dry run to infer all parameter shapes and subsequently initializes the parameters. It will be used later when default random initializations are not desired.

```
@d2l.add_to_class(d2l.Module)  #@save
def apply_init(self, inputs, init=None):
    self.forward(*inputs)
    if init is not None:
        self.net.apply(init)
```

## 6.4.1 Summary

Lazy initialization can be convenient, allowing the framework to infer parameter shapes automatically, making it easy to modify architectures and eliminating one common source of errors. We can pass data through the model to make the framework finally initialize parameters.

## 6.4.2 Exercises

1. What happens if you specify the input dimensions to the first layer but not to subsequent layers? Do you get immediate initialization?
2. What happens if you specify mismatching dimensions?
3. What would you need to do if you have input of varying dimensionality? Hint: look at the parameter tying.

Discussions[114].

# 6.5 Custom Layers

One factor behind deep learning's success is the availability of a wide range of layers that can be composed in creative ways to design architectures suitable for a wide variety of tasks. For instance, researchers have invented layers specifically for handling images, text, looping over sequential data, and performing dynamic programming. Sooner or later, you will need a layer that does not exist yet in the deep learning framework. In these cases, you must build a custom layer. In this section, we show you how.

```
import torch
from torch import nn
from torch.nn import functional as F
from d2l import torch as d2l
```

## 6.5.1 Layers without Parameters

To start, we construct a custom layer that does not have any parameters of its own. This should look familiar if you recall our introduction to modules in Section 6.1. The following `CenteredLayer` class simply subtracts the mean from its input. To build it, we simply need to inherit from the base layer class and implement the forward propagation function.

```
class CenteredLayer(nn.Module):
    def __init__(self):
        super().__init__()

    def forward(self, X):
        return X - X.mean()
```

Let's verify that our layer works as intended by feeding some data through it.

```
layer = CenteredLayer()
layer(torch.tensor([1.0, 2, 3, 4, 5]))
```

```
tensor([-2., -1.,  0.,  1.,  2.])
```

We can now incorporate our layer as a component in constructing more complex models.

```
net = nn.Sequential(nn.LazyLinear(128), CenteredLayer())
```

As an extra sanity check, we can send random data through the network and check that the mean is in fact 0. Because we are dealing with floating point numbers, we may still see a very small nonzero number due to quantization.

```
Y = net(torch.rand(4, 8))
Y.mean()
```

```
tensor(2.3283e-09, grad_fn=<MeanBackward0>)
```

### 6.5.2 Layers with Parameters

Now that we know how to define simple layers, let's move on to defining layers with parameters that can be adjusted through training. We can use built-in functions to create parameters, which provide some basic housekeeping functionality. In particular, they govern access, initialization, sharing, saving, and loading model parameters. This way, among other benefits, we will not need to write custom serialization routines for every custom layer.

Now let's implement our own version of the fully connected layer. Recall that this layer requires two parameters, one to represent the weight and the other for the bias. In this implementation, we bake in the ReLU activation as a default. This layer requires two input arguments: `in_units` and `units`, which denote the number of inputs and outputs, respectively.

```
class MyLinear(nn.Module):
    def __init__(self, in_units, units):
        super().__init__()
        self.weight = nn.Parameter(torch.randn(in_units, units))
        self.bias = nn.Parameter(torch.randn(units,))

    def forward(self, X):
        linear = torch.matmul(X, self.weight.data) + self.bias.data
        return F.relu(linear)
```

Next, we instantiate the `MyLinear` class and access its model parameters.

```
linear = MyLinear(5, 3)
linear.weight
```

```
Parameter containing:
tensor([[-1.3509, -0.6648, -0.3179],
        [-1.0579,  1.0153,  0.9971],
        [ 1.4325, -0.7638,  0.6001],
        [ 0.1169, -1.3587, -0.2305],
        [ 0.0751,  0.4427,  1.0031]], requires_grad=True)
```

We can directly carry out forward propagation calculations using custom layers.

```
linear(torch.rand(2, 5))
```

```
tensor([[0.0000, 0.0000, 2.1424],
        [0.2937, 0.0000, 3.0267]])
```

We can also construct models using custom layers. Once we have that we can use it just like the built-in fully connected layer.

```
net = nn.Sequential(MyLinear(64, 8), MyLinear(8, 1))
net(torch.rand(2, 64))
```

```
tensor([[0.7980],
        [5.1926]])
```

### 6.5.3 Summary

We can design custom layers via the basic layer class. This allows us to define flexible new layers that behave differently from any existing layers in the library. Once defined, custom layers can be invoked in arbitrary contexts and architectures. Layers can have local parameters, which can be created through built-in functions.

### 6.5.4 Exercises

1. Design a layer that takes an input and computes a tensor reduction, i.e., it returns $y_k = \sum_{i,j} W_{ijk} x_i x_j$.
2. Design a layer that returns the leading half of the Fourier coefficients of the data.

Discussions[115].

## 6.6 File I/O

So far we have discussed how to process data and how to build, train, and test deep learning models. However, at some point we will hopefully be happy enough with the learned models that we will want to save the results for later use in various contexts (perhaps even to make predictions in deployment). Additionally, when running a long training process, the best practice is to periodically save intermediate results (checkpointing) to ensure that we do not lose several days' worth of computation if we trip over the power cord of our server. Thus it is time to learn how to load and store both individual weight vectors and entire models. This section addresses both issues.

```
import torch
from torch import nn
from torch.nn import functional as F
```

## 6.6.1 Loading and Saving Tensors

For individual tensors, we can directly invoke the `load` and `save` functions to read and write them respectively. Both functions require that we supply a name, and `save` requires as input the variable to be saved.

```
x = torch.arange(4)
torch.save(x, 'x-file')
```

We can now read the data from the stored file back into memory.

```
x2 = torch.load('x-file')
x2
```

```
tensor([0, 1, 2, 3])
```

We can store a list of tensors and read them back into memory.

```
y = torch.zeros(4)
torch.save([x, y],'x-files')
x2, y2 = torch.load('x-files')
(x2, y2)
```

```
(tensor([0, 1, 2, 3]), tensor([0., 0., 0., 0.]))
```

We can even write and read a dictionary that maps from strings to tensors. This is convenient when we want to read or write all the weights in a model.

```
mydict = {'x': x, 'y': y}
torch.save(mydict, 'mydict')
mydict2 = torch.load('mydict')
mydict2
```

```
{'x': tensor([0, 1, 2, 3]), 'y': tensor([0., 0., 0., 0.])}
```

## 6.6.2 Loading and Saving Model Parameters

Saving individual weight vectors (or other tensors) is useful, but it gets very tedious if we want to save (and later load) an entire model. After all, we might have hundreds of parameter groups sprinkled throughout. For this reason the deep learning framework provides built-in functionalities to load and save entire networks. An important detail to note is that this saves model *parameters* and not the entire model. For example, if we have a 3-layer MLP, we need to specify the architecture separately. The reason for this is that the models themselves can contain arbitrary code, hence they cannot be serialized as naturally. Thus, in order to reinstate a model, we need to generate the architecture in code and then load the parameters from disk. Let's start with our familiar MLP.

```
class MLP(nn.Module):
    def __init__(self):
        super().__init__()
        self.hidden = nn.LazyLinear(256)
        self.output = nn.LazyLinear(10)

    def forward(self, x):
        return self.output(F.relu(self.hidden(x)))

net = MLP()
X = torch.randn(size=(2, 20))
Y = net(X)
```

Next, we store the parameters of the model as a file with the name "mlp.params".

```
torch.save(net.state_dict(), 'mlp.params')
```

To recover the model, we instantiate a clone of the original MLP model. Instead of randomly initializing the model parameters, we read the parameters stored in the file directly.

```
clone = MLP()
clone.load_state_dict(torch.load('mlp.params'))
clone.eval()
```

```
MLP(
  (hidden): LazyLinear(in_features=0, out_features=256, bias=True)
  (output): LazyLinear(in_features=0, out_features=10, bias=True)
)
```

Since both instances have the same model parameters, the computational result of the same input `X` should be the same. Let's verify this.

```
Y_clone = clone(X)
Y_clone == Y
```

```
tensor([[True, True, True, True, True, True, True, True, True, True],
        [True, True, True, True, True, True, True, True, True, True]])
```

## 6.6.3 Summary

The `save` and `load` functions can be used to perform file I/O for tensor objects. We can save and load the entire sets of parameters for a network via a parameter dictionary. Saving the architecture has to be done in code rather than in parameters.

### 6.6.4 Exercises

1. Even if there is no need to deploy trained models to a different device, what are the practical benefits of storing model parameters?
2. Assume that we want to reuse only parts of a network to be incorporated into a network having a different architecture. How would you go about using, say the first two layers from a previous network in a new network?
3. How would you go about saving the network architecture and parameters? What restrictions would you impose on the architecture?

Discussions[116].

## 6.7 GPUs

In Table 1.5.1, we illustrated the rapid growth of computation over the past two decades. In a nutshell, GPU performance has increased by a factor of 1000 every decade since 2000. This offers great opportunities but it also suggests that there was significant demand for such performance.

In this section, we begin to discuss how to harness this computational performance for your research. First by using a single GPU and at a later point, how to use multiple GPUs and multiple servers (with multiple GPUs).

Specifically, we will discuss how to use a single NVIDIA GPU for calculations. First, make sure you have at least one NVIDIA GPU installed. Then, download the NVIDIA driver and CUDA[117] and follow the prompts to set the appropriate path. Once these preparations are complete, the `nvidia-smi` command can be used to view the graphics card information.

In PyTorch, every array has a device; we often refer it as a *context*. So far, by default, all variables and associated computation have been assigned to the CPU. Typically, other contexts might be various GPUs. Things can get even hairier when we deploy jobs across multiple servers. By assigning arrays to contexts intelligently, we can minimize the time spent transferring data between devices. For example, when training neural networks on a server with a GPU, we typically prefer for the model's parameters to live on the GPU.

To run the programs in this section, you need at least two GPUs. Note that this might be extravagant for most desktop computers but it is easily available in the cloud, e.g., by using the AWS EC2 multi-GPU instances. Almost all other sections do *not* require multiple GPUs, but here we simply wish to illustrate data flow between different devices.

```
import torch
from torch import nn
from d2l import torch as d2l
```

## 6.7.1 Computing Devices

We can specify devices, such as CPUs and GPUs, for storage and calculation. By default, tensors are created in the main memory and then the CPU is used for calculations.

In PyTorch, the CPU and GPU can be indicated by `torch.device('cpu')` and `torch.device('cuda')`. It should be noted that the cpu device means all physical CPUs and memory. This means that PyTorch's calculations will try to use all CPU cores. However, a gpu device only represents one card and the corresponding memory. If there are multiple GPUs, we use `torch.device(f'cuda:{i}')` to represent the $i^{\mathrm{th}}$ GPU ($i$ starts at 0). Also, `gpu:0` and gpu are equivalent.

```
def cpu():  #@save
    """Get the CPU device."""
    return torch.device('cpu')

def gpu(i=0):  #@save
    """Get a GPU device."""
    return torch.device(f'cuda:{i}')

cpu(), gpu(), gpu(1)
```

```
(device(type='cpu'),
 device(type='cuda', index=0),
 device(type='cuda', index=1))
```

We can query the number of available GPUs.

```
def num_gpus():  #@save
    """Get the number of available GPUs."""
    return torch.cuda.device_count()

num_gpus()
```

```
2
```

Now we define two convenient functions that allow us to run code even if the requested GPUs do not exist.

```python
def try_gpu(i=0):  #@save
    """Return gpu(i) if exists, otherwise return cpu()."""
    if num_gpus() >= i + 1:
        return gpu(i)
    return cpu()

def try_all_gpus():  #@save
    """Return all available GPUs, or [cpu(),] if no GPU exists."""
    return [gpu(i) for i in range(num_gpus())]

try_gpu(), try_gpu(10), try_all_gpus()
```

```
(device(type='cuda', index=0),
 device(type='cpu'),
 [device(type='cuda', index=0), device(type='cuda', index=1)])
```

## 6.7.2 Tensors and GPUs

By default, tensors are created on the CPU. We can query the device where the tensor is located.

```python
x = torch.tensor([1, 2, 3])
x.device
```

```
device(type='cpu')
```

It is important to note that whenever we want to operate on multiple terms, they need to be on the same device. For instance, if we sum two tensors, we need to make sure that both arguments live on the same device—otherwise the framework would not know where to store the result or even how to decide where to perform the computation.

### Storage on the GPU

There are several ways to store a tensor on the GPU. For example, we can specify a storage device when creating a tensor. Next, we create the tensor variable `X` on the first gpu. The tensor created on a GPU only consumes the memory of this GPU. We can use the `nvidia-smi` command to view GPU memory usage. In general, we need to make sure that we do not create data that exceeds the GPU memory limit.

```python
X = torch.ones(2, 3, device=try_gpu())
X
```

```
tensor([[1., 1., 1.],
        [1., 1., 1.]], device='cuda:0')
```

Assuming that you have at least two GPUs, the following code will create a random tensor, Y, on the second GPU.

```
Y = torch.rand(2, 3, device=try_gpu(1))
Y
```

```
tensor([[0.2746, 0.9378, 0.7462],
        [0.9504, 0.7667, 0.0495]], device='cuda:1')
```

## Copying

If we want to compute X + Y, we need to decide where to perform this operation. For instance, as shown in Fig. 6.7.1, we can transfer X to the second GPU and perform the operation there. *Do not* simply add X and Y, since this will result in an exception. The runtime engine would not know what to do: it cannot find data on the same device and it fails. Since Y lives on the second GPU, we need to move X there before we can add the two.

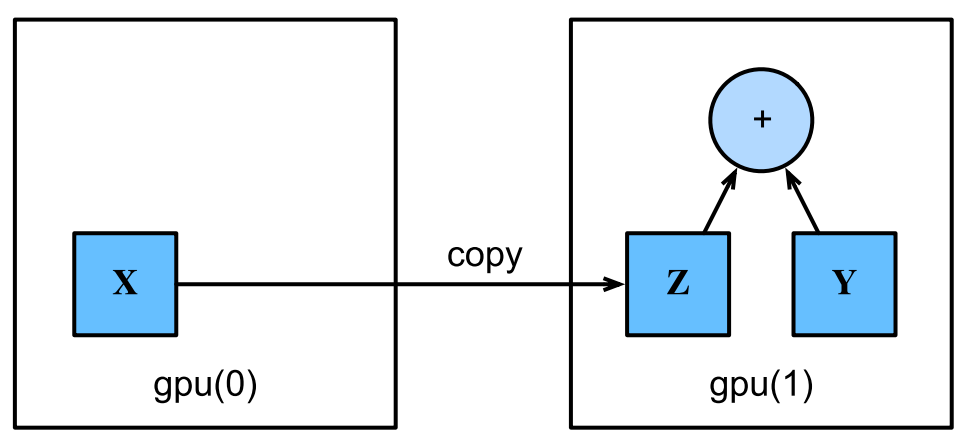

Fig. 6.7.1 Copy data to perform an operation on the same device.

```
Z = X.cuda(1)
print(X)
print(Z)
```

```
tensor([[1., 1., 1.],
        [1., 1., 1.]], device='cuda:0')
tensor([[1., 1., 1.],
        [1., 1., 1.]], device='cuda:1')
```

Now that the data (both Z and Y) are on the same GPU), we can add them up.

```
Y + Z
```

```
tensor([[1.2746, 1.9378, 1.7462],
        [1.9504, 1.7667, 1.0495]], device='cuda:1')
```

But what if your variable Z already lived on your second GPU? What happens if we still call Z.cuda(1)? It will return Z instead of making a copy and allocating new memory.

```
Z.cuda(1) is Z
```

```
True
```

### Side Notes

People use GPUs to do machine learning because they expect them to be fast. But transferring variables between devices is slow: much slower than computation. So we want you to be 100% certain that you want to do something slow before we let you do it. If the deep learning framework just did the copy automatically without crashing then you might not realize that you had written some slow code.

Transferring data is not only slow, it also makes parallelization a lot more difficult, since we have to wait for data to be sent (or rather to be received) before we can proceed with more operations. This is why copy operations should be taken with great care. As a rule of thumb, many small operations are much worse than one big operation. Moreover, several operations at a time are much better than many single operations interspersed in the code unless you know what you are doing. This is the case since such operations can block if one device has to wait for the other before it can do something else. It is a bit like ordering your coffee in a queue rather than pre-ordering it by phone and finding out that it is ready when you are.

Last, when we print tensors or convert tensors to the NumPy format, if the data is not in the main memory, the framework will copy it to the main memory first, resulting in additional transmission overhead. Even worse, it is now subject to the dreaded global interpreter lock that makes everything wait for Python to complete.

## 6.7.3 Neural Networks and GPUs

Similarly, a neural network model can specify devices. The following code puts the model parameters on the GPU.

```
net = nn.Sequential(nn.LazyLinear(1))
net = net.to(device=try_gpu())
```

We will see many more examples of how to run models on GPUs in the following chapters, simply because the models will become somewhat more computationally intensive.

For example, when the input is a tensor on the GPU, the model will calculate the result on the same GPU.

```
net(X)
```

```
tensor([[0.6385],
        [0.6385]], device='cuda:0', grad_fn=<AddmmBackward0>)
```

Let's confirm that the model parameters are stored on the same GPU.

```
net[0].weight.data.device
```

```
device(type='cuda', index=0)
```

Let the trainer support GPU.

```
@d2l.add_to_class(d2l.Trainer)  #@save
def __init__(self, max_epochs, num_gpus=0, gradient_clip_val=0):
    self.save_hyperparameters()
    self.gpus = [d2l.gpu(i) for i in range(min(num_gpus, d2l.num_gpus()))]

@d2l.add_to_class(d2l.Trainer)  #@save
def prepare_batch(self, batch):
    if self.gpus:
        batch = [a.to(self.gpus[0]) for a in batch]
    return batch

@d2l.add_to_class(d2l.Trainer)  #@save
def prepare_model(self, model):
    model.trainer = self
    model.board.xlim = [0, self.max_epochs]
    if self.gpus:
        model.to(self.gpus[0])
    self.model = model
```

In short, as long as all data and parameters are on the same device, we can learn models efficiently. In the following chapters we will see several such examples.

## 6.7.4 Summary

We can specify devices for storage and calculation, such as the CPU or GPU. By default, data is created in the main memory and then uses the CPU for calculations. The deep learning framework requires all input data for calculation to be on the same device, be it CPU or the same GPU. You can lose significant performance by moving data without care. A typical mistake is as follows: computing the loss for every minibatch on the GPU and reporting it back to the user on the command line (or logging it in a NumPy `ndarray`) will trigger a global interpreter lock which stalls all GPUs. It is much better to allocate memory for logging inside the GPU and only move larger logs.

## 6.7.5 Exercises

1. Try a larger computation task, such as the multiplication of large matrices, and see the difference in speed between the CPU and GPU. What about a task with a small number of calculations?
2. How should we read and write model parameters on the GPU?
3. Measure the time it takes to compute 1000 matrix–matrix multiplications of $100 \times 100$ matrices and log the Frobenius norm of the output matrix one result at a time. Compare it with keeping a log on the GPU and transferring only the final result.
4. Measure how much time it takes to perform two matrix–matrix multiplications on two GPUs at the same time. Compare it with computing in in sequence on one GPU. Hint: you should see almost linear scaling.

Discussions[118].

# 7 Convolutional Neural Networks

Image data is represented as a two-dimensional grid of pixels, be the image monochromatic or in color. Accordingly each pixel corresponds to one or multiple numerical values respectively. So far we have ignored this rich structure and treated images as vectors of numbers by *flattening* them, irrespective of the spatial relation between pixels. This deeply unsatisfying approach was necessary in order to feed the resulting one-dimensional vectors through a fully connected MLP.

Because these networks are invariant to the order of the features, we could get similar results regardless of whether we preserve an order corresponding to the spatial structure of the pixels or if we permute the columns of our design matrix before fitting the MLP's parameters. Ideally, we would leverage our prior knowledge that nearby pixels are typically related to each other, to build efficient models for learning from image data.

This chapter introduces *convolutional neural networks* (CNNs) (LeCun *et al.*, 1995), a powerful family of neural networks that are designed for precisely this purpose. CNN-based architectures are now ubiquitous in the field of computer vision. For instance, on the Imagnet collection (Deng *et al.*, 2009) it was only the use of convolutional neural networks, in short Convnets, that provided significant performance improvements (Krizhevsky *et al.*, 2012).

Modern CNNs, as they are called colloquially, owe their design to inspirations from biology, group theory, and a healthy dose of experimental tinkering. In addition to their sample efficiency in achieving accurate models, CNNs tend to be computationally efficient, both because they require fewer parameters than fully connected architectures and because convolutions are easy to parallelize across GPU cores (Chetlur *et al.*, 2014). Consequently, practitioners often apply CNNs whenever possible, and increasingly they have emerged as credible competitors even on tasks with a one-dimensional sequence structure, such as audio (Abdel-Hamid *et al.*, 2014), text (Kalchbrenner *et al.*, 2014), and time series analysis (LeCun *et al.*, 1995), where recurrent neural networks are conventionally used. Some clever adaptations of CNNs have also brought them to bear on graph-structured data (Kipf and Welling, 2016) and in recommender systems.

First, we will dive more deeply into the motivation for convolutional neural networks. This is followed by a walk through the basic operations that comprise the backbone of all convolutional networks. These include the convolutional layers themselves, nitty-gritty details including padding and stride, the pooling layers used to aggregate information across adjacent spatial regions, the use of multiple channels at each layer, and a careful discussion of the structure of modern architectures. We will conclude the chapter with a full working ex-ample of LeNet, the first convolutional network successfully deployed, long before the rise of modern deep learning. In the next chapter, we will dive into full implementations of some popular and comparatively recent CNN architectures whose designs represent most of the techniques commonly used by modern practitioners.

## 7.1 From Fully Connected Layers to Convolutions

To this day, the models that we have discussed so far remain appropriate options when we are dealing with tabular data. By tabular, we mean that the data consist of rows corresponding to examples and columns corresponding to features. With tabular data, we might anticipate that the patterns we seek could involve interactions among the features, but we do not assume any structure *a priori* concerning how the features interact.

Sometimes, we truly lack the knowledge to be able to guide the construction of fancier architectures. In these cases, an MLP may be the best that we can do. However, for high-dimensional perceptual data, such structureless networks can grow unwieldy.

For instance, let's return to our running example of distinguishing cats from dogs. Say that we do a thorough job in data collection, collecting an annotated dataset of one-megapixel photographs. This means that each input to the network has one million dimensions. Even an aggressive reduction to one thousand hidden dimensions would require a fully connected layer characterized by $10^6 \times 10^3 = 10^9$ parameters. Unless we have lots of GPUs, a talent for distributed optimization, and an extraordinary amount of patience, learning the parameters of this network may turn out to be infeasible.

A careful reader might object to this argument on the basis that one megapixel resolution may not be necessary. However, while we might be able to get away with one hundred thousand pixels, our hidden layer of size 1000 grossly underestimates the number of hidden units that it takes to learn good representations of images, so a practical system will still require billions of parameters. Moreover, learning a classifier by fitting so many parameters might require collecting an enormous dataset. And yet today both humans and computers are able to distinguish cats from dogs quite well, seemingly contradicting these intuitions. That is because images exhibit rich structure that can be exploited by humans and machine learning models alike. Convolutional neural networks (CNNs) are one creative way that machine learning has embraced for exploiting some of the known structure in natural images.

### 7.1.1 Invariance

Imagine that we want to detect an object in an image. It seems reasonable that whatever method we use to recognize objects should not be overly concerned with the precise location of the object in the image. Ideally, our system should exploit this knowledge. Pigs usually do not fly and planes usually do not swim. Nonetheless, we should still recognize a pig were one to appear at the top of the image. We can draw some inspiration here from the children's game "Where's Waldo" (which itself has inspired many real-life imitations, such as that depicted in Fig. 7.1.1). The game consists of a number of chaotic scenes bursting with activities. Waldo shows up somewhere in each, typically lurking in some unlikely location. The reader's goal is to locate him. Despite his characteristic outfit, this can be surprisingly difficult, due to the large number of distractions. However, *what Waldo looks like* does not depend upon *where Waldo is located*. We could sweep the image with a Waldo detector that could assign a score to each patch, indicating the likelihood that the patch contains Waldo. In fact, many object detection and segmentation algorithms are based on this approach (Long *et al.*, 2015). CNNs systematize this idea of *spatial invariance*, exploiting it to learn useful representations with fewer parameters.

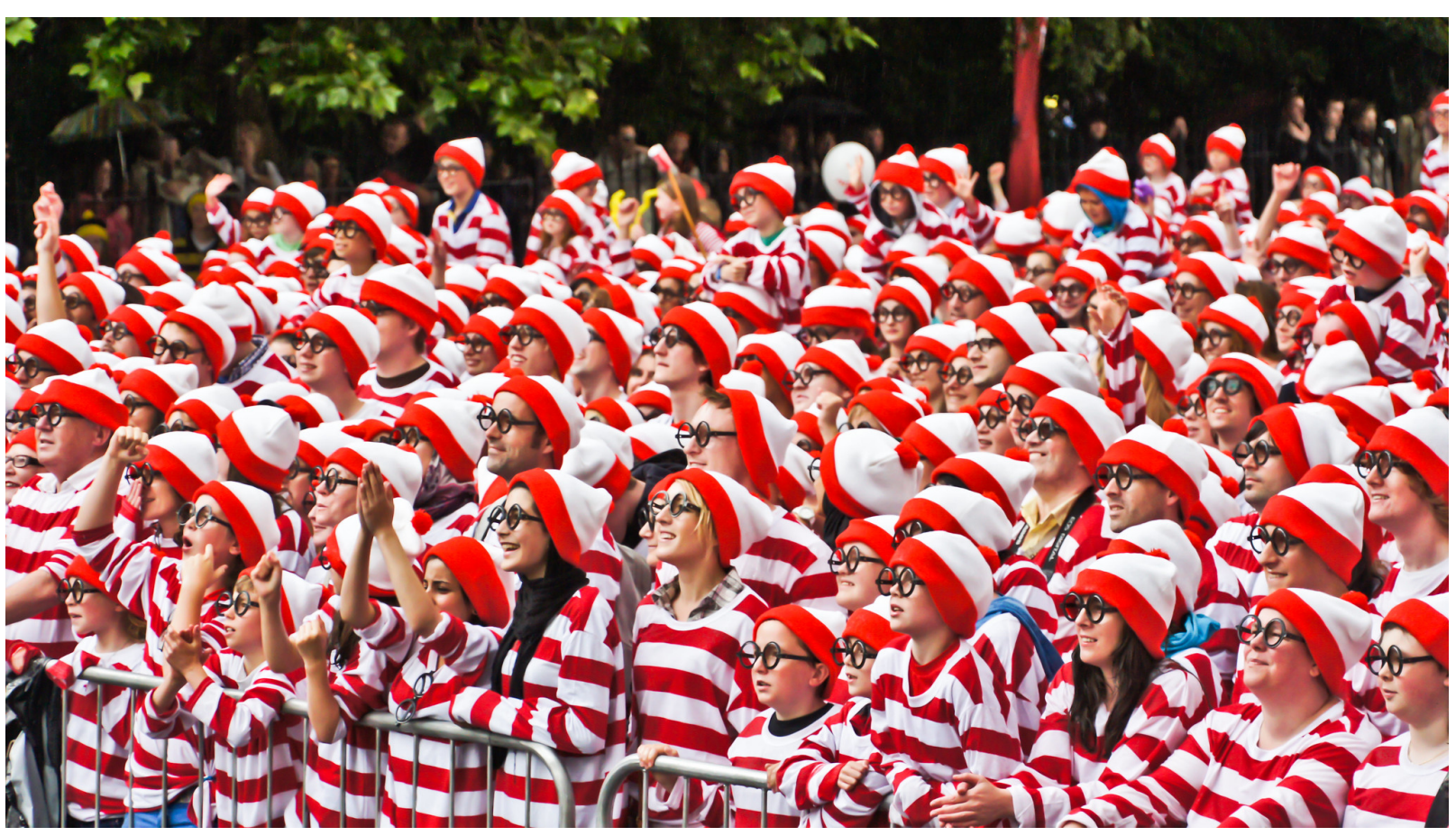

Fig. 7.1.1 Can you find Waldo (image courtesy of William Murphy (Infomatique))?

We can now make these intuitions more concrete by enumerating a few desiderata to guide our design of a neural network architecture suitable for computer vision:

1. In the earliest layers, our network should respond similarly to the same patch, regardless of where it appears in the image. This principle is called *translation invariance* (or *translation equivariance*).

2. The earliest layers of the network should focus on local regions, without regard for the contents of the image in distant regions. This is the *locality* principle. Eventually, these local representations can be aggregated to make predictions at the whole image level.

3. As we proceed, deeper layers should be able to capture longer-range features of the image, in a way similar to higher level vision in nature.

Let's see how this translates into mathematics.

### 7.1.2 Constraining the MLP

To start off, we can consider an MLP with two-dimensional images $\mathbf{X}$ as inputs and their immediate hidden representations $\mathbf{H}$ similarly represented as matrices (they are two-dimensional tensors in code), where both $\mathbf{X}$ and $\mathbf{H}$ have the same shape. Let that sink in. We now imagine that not only the inputs but also the hidden representations possess spatial structure.

Let $[\mathbf{X}]_{i,j}$ and $[\mathbf{H}]_{i,j}$ denote the pixel at location $(i, j)$ in the input image and hidden representation, respectively. Consequently, to have each of the hidden units receive input from each of the input pixels, we would switch from using weight matrices (as we did previously in MLPs) to representing our parameters as fourth-order weight tensors $\mathsf{W}$. Suppose that $\mathbf{U}$ contains biases, we could formally express the fully connected layer as

$$\begin{aligned}[\mathbf{H}]_{i,j} &= [\mathbf{U}]_{i,j} + \sum_k \sum_l [\mathsf{W}]_{i,j,k,l} [\mathbf{X}]_{k,l}\\ &= [\mathbf{U}]_{i,j} + \sum_a \sum_b [\mathsf{V}]_{i,j,a,b} [\mathbf{X}]_{i+a,j+b}.\end{aligned} \tag{7.1.1}$$

The switch from $\mathsf{W}$ to $\mathsf{V}$ is entirely cosmetic for now since there is a one-to-one correspondence between coefficients in both fourth-order tensors. We simply re-index the subscripts $(k, l)$ such that $k = i + a$ and $l = j + b$. In other words, we set $[\mathsf{V}]_{i,j,a,b} = [\mathsf{W}]_{i,j,i+a,j+b}$. The indices $a$ and $b$ run over both positive and negative offsets, covering the entire image. For any given location $(i, j)$ in the hidden representation $[\mathbf{H}]_{i,j}$, we compute its value by summing over pixels in $x$, centered around $(i, j)$ and weighted by $[\mathsf{V}]_{i,j,a,b}$. Before we carry on, let's consider the total number of parameters required for a *single* layer in this parametrization: a $1000 \times 1000$ image (1 megapixel) is mapped to a $1000 \times 1000$ hidden representation. This requires $10^{12}$ parameters, far beyond what computers currently can handle.

## Translation Invariance

Now let's invoke the first principle established above: translation invariance (Zhang *et al.*, 1988). This implies that a shift in the input $\mathbf{X}$ should simply lead to a shift in the hidden representation $\mathbf{H}$. This is only possible if $\mathsf{V}$ and $\mathbf{U}$ do not actually depend on $(i, j)$. As such, we have $[\mathsf{V}]_{i,j,a,b} = [\mathbf{V}]_{a,b}$ and $\mathbf{U}$ is a constant, say $u$. As a result, we can simplify the definition for $\mathbf{H}$:

$$[\mathbf{H}]_{i,j} = u + \sum_a \sum_b [\mathbf{V}]_{a,b} [\mathbf{X}]_{i+a,j+b}. \tag{7.1.2}$$

This is a *convolution*! We are effectively weighting pixels at $(i + a, j + b)$ in the vicinity of location $(i, j)$ with coefficients $[\mathbf{V}]_{a,b}$ to obtain the value $[\mathbf{H}]_{i,j}$. Note that $[\mathbf{V}]_{a,b}$ needs many fewer coefficients than $[\mathsf{V}]_{i,j,a,b}$ since it no longer depends on the location within the image. Consequently, the number of parameters required is no longer $10^{12}$ but a much more reasonable $4 \times 10^6$: we still have the dependency on $a, b \in (-1000, 1000)$. In short, we have made significant progress. Time-delay neural networks (TDNNs) are some of the first examples to exploit this idea (Waibel *et al.*, 1989).

### Locality

Now let's invoke the second principle: locality. As motivated above, we believe that we should not have to look very far away from location $(i, j)$ in order to glean relevant information to assess what is going on at $[\mathbf{H}]_{i,j}$. This means that outside some range $|a| > \Delta$ or $|b| > \Delta$, we should set $[\mathbf{V}]_{a,b} = 0$. Equivalently, we can rewrite $[\mathbf{H}]_{i,j}$ as

$$[\mathbf{H}]_{i,j} = u + \sum_{a=-\Delta}^{\Delta} \sum_{b=-\Delta}^{\Delta} [\mathbf{V}]_{a,b} [\mathbf{X}]_{i+a,j+b}. \tag{7.1.3}$$

This reduces the number of parameters from $4 \times 10^6$ to $4\Delta^2$, where $\Delta$ is typically smaller than 10. As such, we reduced the number of parameters by another four orders of magnitude. Note that (7.1.3), is what is called, in a nutshell, a *convolutional layer*. *Convolutional neural networks* (CNNs) are a special family of neural networks that contain convolutional layers. In the deep learning research community, $\mathbf{V}$ is referred to as a *convolution kernel*, a *filter*, or simply the layer's *weights* that are learnable parameters.

While previously, we might have required billions of parameters to represent just a single layer in an image-processing network, we now typically need just a few hundred, without altering the dimensionality of either the inputs or the hidden representations. The price paid for this drastic reduction in parameters is that our features are now translation invariant and that our layer can only incorporate local information, when determining the value of each hidden activation. All learning depends on imposing inductive bias. When that bias agrees with reality, we get sample-efficient models that generalize well to unseen data. But of course, if those biases do not agree with reality, e.g., if images turned out not to be translation invariant, our models might struggle even to fit our training data.

This dramatic reduction in parameters brings us to our last desideratum, namely that deeper layers should represent larger and more complex aspects of an image. This can be achieved by interleaving nonlinearities and convolutional layers repeatedly.

## 7.1.3 Convolutions

Let's briefly review why (7.1.3) is called a convolution. In mathematics, the *convolution* between two functions (Rudin, 1973), say $f, g : \mathbb{R}^d \to \mathbb{R}$ is defined as

$$(f * g)(\mathbf{x}) = \int f(\mathbf{z}) g(\mathbf{x} - \mathbf{z}) d\mathbf{z}. \tag{7.1.4}$$

That is, we measure the overlap between $f$ and $g$ when one function is "flipped" and shifted by $\mathbf{x}$. Whenever we have discrete objects, the integral turns into a sum. For instance, for vectors from the set of square-summable infinite-dimensional vectors with index running over $\mathbb{Z}$ we obtain the following definition:

$$(f * g)(i) = \sum_a f(a) g(i - a). \tag{7.1.5}$$

For two-dimensional tensors, we have a corresponding sum with indices $(a, b)$ for $f$ and $(i - a, j - b)$ for $g$, respectively:

$$(f * g)(i, j) = \sum_a \sum_b f(a, b) g(i - a, j - b). \tag{7.1.6}$$

This looks similar to (7.1.3), with one major difference. Rather than using $(i + a, j + b)$, we are using the difference instead. Note, though, that this distinction is mostly cosmetic since we can always match the notation between (7.1.3) and (7.1.6). Our original definition in (7.1.3) more properly describes a *cross-correlation*. We will come back to this in the following section.

## 7.1.4 Channels

Returning to our Waldo detector, let's see what this looks like. The convolutional layer picks windows of a given size and weighs intensities according to the filter $\mathsf{V}$, as demonstrated in Fig. 7.1.2. We might aim to learn a model so that wherever the "waldoness" is highest, we should find a peak in the hidden layer representations.

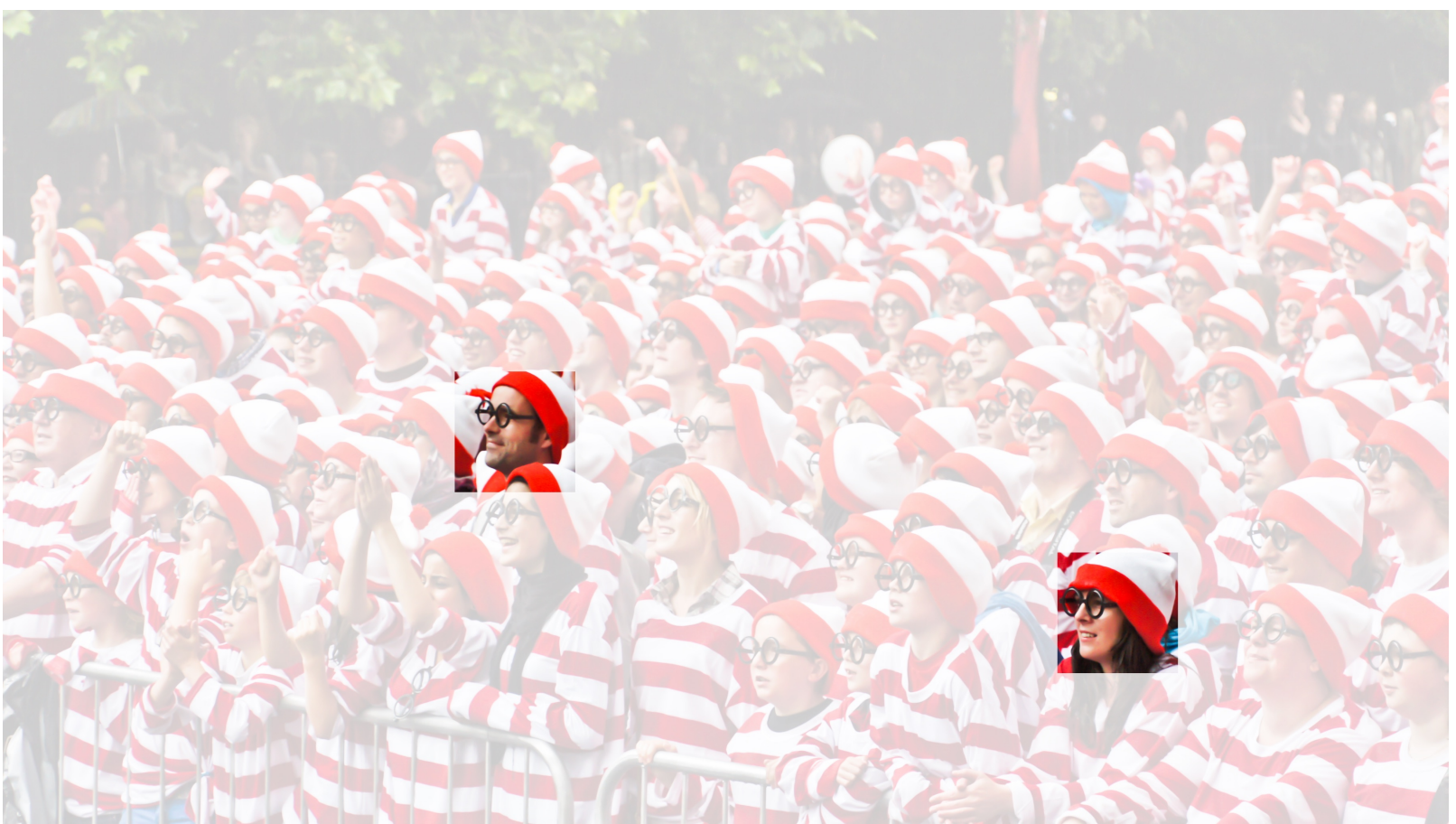

Fig. 7.1.2 Detect Waldo (image courtesy of William Murphy (Infomatique)).

There is just one problem with this approach. So far, we blissfully ignored that images consist of three channels: red, green, and blue. In sum, images are not two-dimensional objects but rather third-order tensors, characterized by a height, width, and channel, e.g., with shape $1024 \times 1024 \times 3$ pixels. While the first two of these axes concern spatial relationships, the third can be regarded as assigning a multidimensional representation to each pixel location. We thus index $\mathsf{X}$ as $[\mathsf{X}]_{i,j,k}$. The convolutional filter has to adapt accordingly. Instead of $[\mathbf{V}]_{a,b}$, we now have $[\mathsf{V}]_{a,b,c}$.

Moreover, just as our input consists of a third-order tensor, it turns out to be a good idea to similarly formulate our hidden representations as third-order tensors $\mathsf{H}$. In other words, rather than just having a single hidden representation corresponding to each spatial location, we want an entire vector of hidden representations corresponding to each spatial location.

We could think of the hidden representations as comprising a number of two-dimensional grids stacked on top of each other. As in the inputs, these are sometimes called *channels*. They are also sometimes called *feature maps*, as each provides a spatialized set of learned features for the subsequent layer. Intuitively, you might imagine that at lower layers that are closer to inputs, some channels could become specialized to recognize edges while others could recognize textures.

To support multiple channels in both inputs ($\mathsf{X}$) and hidden representations ($\mathsf{H}$), we can add a fourth coordinate to $\mathsf{V}$: $[\mathsf{V}]_{a,b,c,d}$. Putting everything together we have:

$$[\mathsf{H}]_{i,j,d} = \sum_{a=-\Delta}^{\Delta} \sum_{b=-\Delta}^{\Delta} \sum_{c} [\mathsf{V}]_{a,b,c,d} [\mathsf{X}]_{i+a,j+b,c}, \tag{7.1.7}$$

where $d$ indexes the output channels in the hidden representations $\mathsf{H}$. The subsequent convolutional layer will go on to take a third-order tensor, $\mathsf{H}$, as input. We take (7.1.7), because of its generality, as the definition of a convolutional layer for multiple channels, where $\mathsf{V}$ is a kernel or filter of the layer.

There are still many operations that we need to address. For instance, we need to figure out how to combine all the hidden representations to a single output, e.g., whether there is a Waldo *anywhere* in the image. We also need to decide how to compute things efficiently, how to combine multiple layers, appropriate activation functions, and how to make reasonable design choices to yield networks that are effective in practice. We turn to these issues in the remainder of the chapter.

### 7.1.5 Summary and Discussion

In this section we derived the structure of convolutional neural networks from first principles. While it is unclear whether this was the route taken to the invention of CNNs, it is satisfying to know that they are the *right* choice when applying reasonable principles to how image processing and computer vision algorithms should operate, at least at lower levels. In particular, translation invariance in images implies that all patches of an image will be treated in the same manner. Locality means that only a small neighborhood of pixels will be used to compute the corresponding hidden representations. Some of the earliest references to CNNs are in the form of the Neocognitron (Fukushima, 1982).

A second principle that we encountered in our reasoning is how to reduce the number of parameters in a function class without limiting its expressive power, at least, whenever certain assumptions on the model hold. We saw a dramatic reduction of complexity as a result of this restriction, turning computationally and statistically infeasible problems into tractable models.

Adding channels allowed us to bring back some of the complexity that was lost due to the restrictions imposed on the convolutional kernel by locality and translation invariance. Note that it is quite natural to add channels other than just red, green, and blue. Many satellite images, in particular for agriculture and meteorology, have tens to hundreds of channels, generating hyperspectral images instead. They report data on many different wavelengths.

In the following we will see how to use convolutions effectively to manipulate the dimensionality of the images they operate on, how to move from location-based to channel-based representations, and how to deal with large numbers of categories efficiently.

### 7.1.6 Exercises

1. Assume that the size of the convolution kernel is $\Delta = 0$. Show that in this case the convolution kernel implements an MLP independently for each set of channels. This leads to the Network in Network architectures (Lin *et al.*, 2013).
2. Audio data is often represented as a one-dimensional sequence.
   1. When might you want to impose locality and translation invariance for audio?
   2. Derive the convolution operations for audio.
   3. Can you treat audio using the same tools as computer vision? Hint: use the spectrogram.
3. Why might translation invariance not be a good idea after all? Give an example.
4. Do you think that convolutional layers might also be applicable for text data? Which problems might you encounter with language?
5. What happens with convolutions when an object is at the boundary of an image?
6. Prove that the convolution is symmetric, i.e., $f * g = g * f$.

Discussions[119].

## 7.2 Convolutions for Images

Now that we understand how convolutional layers work in theory, we are ready to see how they work in practice. Building on our motivation of convolutional neural networks as efficient architectures for exploring structure in image data, we stick with images as our running example.

```
import torch
from torch import nn
from d2l import torch as d2l
```

### 7.2.1 The Cross-Correlation Operation

Recall that strictly speaking, convolutional layers are a misnomer, since the operations they express are more accurately described as cross-correlations. Based on our descriptions of convolutional layers in Section 7.1, in such a layer, an input tensor and a kernel tensor are combined to produce an output tensor through a cross-correlation operation.

Let's ignore channels for now and see how this works with two-dimensional data and hidden representations. In Fig. 7.2.1, the input is a two-dimensional tensor with a height of 3 and width of 3. We mark the shape of the tensor as $3 \times 3$ or $(3, 3)$. The height and width of the kernel are both 2. The shape of the *kernel window* (or *convolution window*) is given by the height and width of the kernel (here it is $2 \times 2$).

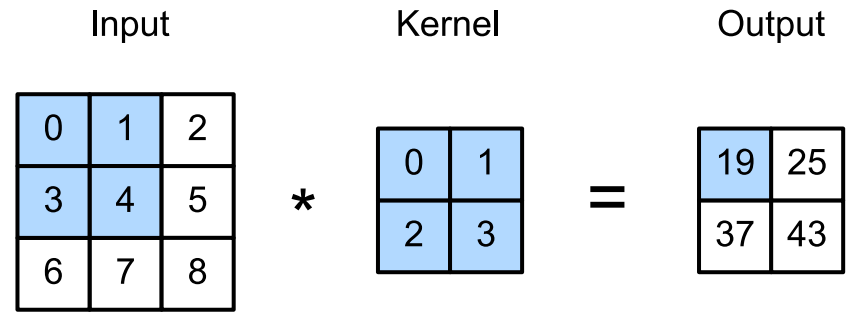

Fig. 7.2.1 Two-dimensional cross-correlation operation. The shaded portions are the first output element as well as the input and kernel tensor elements used for the output computation: $0 \times 0 + 1 \times 1 + 3 \times 2 + 4 \times 3 = 19$.

In the two-dimensional cross-correlation operation, we begin with the convolution window positioned at the upper-left corner of the input tensor and slide it across the input tensor, both from left to right and top to bottom. When the convolution window slides to a certain position, the input subtensor contained in that window and the kernel tensor are multiplied elementwise and the resulting tensor is summed up yielding a single scalar value. This result gives the value of the output tensor at the corresponding location. Here, the output tensor has a height of 2 and width of 2 and the four elements are derived from the two-dimensional cross-correlation operation:

$$\begin{aligned}
0 \times 0 + 1 \times 1 + 3 \times 2 + 4 \times 3 &= 19,\\
1 \times 0 + 2 \times 1 + 4 \times 2 + 5 \times 3 &= 25,\\
3 \times 0 + 4 \times 1 + 6 \times 2 + 7 \times 3 &= 37,\\
4 \times 0 + 5 \times 1 + 7 \times 2 + 8 \times 3 &= 43.
\end{aligned} \tag{7.2.1}$$

Note that along each axis, the output size is slightly smaller than the input size. Because the kernel has width and height greater than 1, we can only properly compute the cross-correlation for locations where the kernel fits wholly within the image, the output size is given by the input size $n_\textrm{h} \times n_\textrm{w}$ minus the size of the convolution kernel $k_\textrm{h} \times k_\textrm{w}$ via

$$(n_\textrm{h}-k_\textrm{h}+1) \times (n_\textrm{w}-k_\textrm{w}+1). \tag{7.2.2}$$

This is the case since we need enough space to "shift" the convolution kernel across the image. Later we will see how to keep the size unchanged by padding the image with zeros around its boundary so that there is enough space to shift the kernel. Next, we implement this process in the `corr2d` function, which accepts an input tensor `X` and a kernel tensor `K` and returns an output tensor `Y`.

```
def corr2d(X, K):  #@save
    """Compute 2D cross-correlation."""
    h, w = K.shape
    Y = torch.zeros((X.shape[0] - h + 1, X.shape[1] - w + 1))
    for i in range(Y.shape[0]):
        for j in range(Y.shape[1]):
            Y[i, j] = (X[i:i + h, j:j + w] * K).sum()
    return Y
```

We can construct the input tensor `X` and the kernel tensor `K` from Fig. 7.2.1 to validate the output of the above implementation of the two-dimensional cross-correlation operation.

```
X = torch.tensor([[0.0, 1.0, 2.0], [3.0, 4.0, 5.0], [6.0, 7.0, 8.0]])
K = torch.tensor([[0.0, 1.0], [2.0, 3.0]])
corr2d(X, K)
```

```
tensor([[19., 25.],
        [37., 43.]])
```

## 7.2.2 Convolutional Layers

A convolutional layer cross-correlates the input and kernel and adds a scalar bias to produce an output. The two parameters of a convolutional layer are the kernel and the scalar bias. When training models based on convolutional layers, we typically initialize the kernels randomly, just as we would with a fully connected layer.

We are now ready to implement a two-dimensional convolutional layer based on the `corr2d` function defined above. In the `__init__` constructor method, we declare `weight` and `bias` as the two model parameters. The forward propagation method calls the `corr2d` function and adds the bias.

```
class Conv2D(nn.Module):
    def __init__(self, kernel_size):
        super().__init__()
        self.weight = nn.Parameter(torch.rand(kernel_size))
        self.bias = nn.Parameter(torch.zeros(1))

    def forward(self, x):
        return corr2d(x, self.weight) + self.bias
```

In $h \times w$ convolution or an $h \times w$ convolution kernel, the height and width of the convolution kernel are $h$ and $w$, respectively. We also refer to a convolutional layer with an $h \times w$ convolution kernel simply as an $h \times w$ convolutional layer.

## 7.2.3 Object Edge Detection in Images

Let's take a moment to parse a simple application of a convolutional layer: detecting the edge of an object in an image by finding the location of the pixel change. First, we construct an "image" of $6 \times 8$ pixels. The middle four columns are black (0) and the rest are white (1).

```
X = torch.ones((6, 8))
X[:, 2:6] = 0
X
```

```
tensor([[1., 1., 0., 0., 0., 0., 1., 1.],
        [1., 1., 0., 0., 0., 0., 1., 1.],
        [1., 1., 0., 0., 0., 0., 1., 1.],
        [1., 1., 0., 0., 0., 0., 1., 1.],
        [1., 1., 0., 0., 0., 0., 1., 1.],
        [1., 1., 0., 0., 0., 0., 1., 1.]])
```

Next, we construct a kernel `K` with a height of 1 and a width of 2. When we perform the cross-correlation operation with the input, if the horizontally adjacent elements are the same, the output is 0. Otherwise, the output is nonzero. Note that this kernel is a special case of a finite difference operator. At location $(i, j)$ it computes $x_{i,j} - x_{(i+1),j}$, i.e., it computes the difference between the values of horizontally adjacent pixels. This is a discrete approximation of the first derivative in the horizontal direction. After all, for a function $f(i, j)$ its derivative $-\partial_i f(i, j) = \lim_{\epsilon \to 0} \frac{f(i,j) - f(i+\epsilon,j)}{\epsilon}$. Let's see how this works in practice.

```
K = torch.tensor([[1.0, -1.0]])
```

We are ready to perform the cross-correlation operation with arguments `X` (our input) and `K` (our kernel). As you can see, we detect 1 for the edge from white to black and $-1$ for the edge from black to white. All other outputs take value 0.

```
Y = corr2d(X, K)
Y
```

```
tensor([[ 0.,  1.,  0.,  0.,  0., -1.,  0.],
        [ 0.,  1.,  0.,  0.,  0., -1.,  0.],
        [ 0.,  1.,  0.,  0.,  0., -1.,  0.],
        [ 0.,  1.,  0.,  0.,  0., -1.,  0.],
        [ 0.,  1.,  0.,  0.,  0., -1.,  0.],
        [ 0.,  1.,  0.,  0.,  0., -1.,  0.]])
```

We can now apply the kernel to the transposed image. As expected, it vanishes. The kernel `K` only detects vertical edges.

```
corr2d(X.t(), K)
```

```
tensor([[0., 0., 0., 0., 0.],
        [0., 0., 0., 0., 0.],
        [0., 0., 0., 0., 0.],
        [0., 0., 0., 0., 0.],
        [0., 0., 0., 0., 0.],
        [0., 0., 0., 0., 0.],
        [0., 0., 0., 0., 0.],
        [0., 0., 0., 0., 0.]])
```

### 7.2.4 Learning a Kernel

Designing an edge detector by finite differences `[1, -1]` is neat if we know this is precisely what we are looking for. However, as we look at larger kernels, and consider successive layers of convolutions, it might be impossible to specify precisely what each filter should be doing manually.

Now let's see whether we can learn the kernel that generated `Y` from `X` by looking at the input–output pairs only. We first construct a convolutional layer and initialize its kernel as a random tensor. Next, in each iteration, we will use the squared error to compare `Y` with the output of the convolutional layer. We can then calculate the gradient to update the kernel. For the sake of simplicity, in the following we use the built-in class for two-dimensional convolutional layers and ignore the bias.

```
# Construct a two-dimensional convolutional layer with 1 output channel and a
# kernel of shape (1, 2). For the sake of simplicity, we ignore the bias here
conv2d = nn.LazyConv2d(1, kernel_size=(1, 2), bias=False)

# The two-dimensional convolutional layer uses four-dimensional input and
# output in the format of (example, channel, height, width), where the batch
# size (number of examples in the batch) and the number of channels are both 1
X = X.reshape((1, 1, 6, 8))
Y = Y.reshape((1, 1, 6, 7))
lr = 3e-2  # Learning rate

for i in range(10):
    Y_hat = conv2d(X)
    l = (Y_hat - Y) ** 2
    conv2d.zero_grad()
    l.sum().backward()
    # Update the kernel
    conv2d.weight.data[:] -= lr * conv2d.weight.grad
    if (i + 1) % 2 == 0:
        print(f'epoch {i + 1}, loss {l.sum():.3f}')
```

```
epoch 2, loss 13.904
epoch 4, loss 2.333
epoch 6, loss 0.392
epoch 8, loss 0.066
epoch 10, loss 0.011
```

Note that the error has dropped to a small value after 10 iterations. Now we will take a look at the kernel tensor we learned.

```
conv2d.weight.data.reshape((1, 2))
```

```
tensor([[ 0.9797, -0.9816]])
```

Indeed, the learned kernel tensor is remarkably close to the kernel tensor `K` we defined earlier.

## 7.2.5 Cross-Correlation and Convolution

Recall our observation from Section 7.1 of the correspondence between the cross-correlation and convolution operations. Here let's continue to consider two-dimensional convolutional layers. What if such layers perform strict convolution operations as defined in (7.1.6) instead of cross-correlations? In order to obtain the output of the strict *convolution* operation, we only need to flip the two-dimensional kernel tensor both horizontally and vertically, and then perform the *cross-correlation* operation with the input tensor.

It is noteworthy that since kernels are learned from data in deep learning, the outputs of convolutional layers remain unaffected no matter such layers perform either the strict convolution operations or the cross-correlation operations.

To illustrate this, suppose that a convolutional layer performs *cross-correlation* and learns the kernel in Fig. 7.2.1, which is here denoted as the matrix $\mathbf{K}$. Assuming that other conditions remain unchanged, when this layer instead performs strict *convolution*, the learned kernel $\mathbf{K}'$ will be the same as $\mathbf{K}$ after $\mathbf{K}'$ is flipped both horizontally and vertically. That is to say, when the convolutional layer performs strict *convolution* for the input in Fig. 7.2.1 and $\mathbf{K}'$, the same output in Fig. 7.2.1 (cross-correlation of the input and $\mathbf{K}$) will be obtained.

In keeping with standard terminology in deep learning literature, we will continue to refer to the cross-correlation operation as a convolution even though, strictly-speaking, it is slightly different. Furthermore, we use the term *element* to refer to an entry (or component) of any tensor representing a layer representation or a convolution kernel.

## 7.2.6 Feature Map and Receptive Field

As described in Section 7.1.4, the convolutional layer output in Fig. 7.2.1 is sometimes called a *feature map*, as it can be regarded as the learned representations (features) in the spatial dimensions (e.g., width and height) to the subsequent layer. In CNNs, for any element $x$ of some layer, its *receptive field* refers to all the elements (from all the previous layers) that may affect the calculation of $x$ during the forward propagation. Note that the receptive field may be larger than the actual size of the input.

Let's continue to use Fig. 7.2.1 to explain the receptive field. Given the $2 \times 2$ convolution kernel, the receptive field of the shaded output element (of value 19) is the four elements in the shaded portion of the input. Now let's denote the $2 \times 2$ output as $\mathbf{Y}$ and consider a deeper CNN with an additional $2 \times 2$ convolutional layer that takes $\mathbf{Y}$ as its input, outputting a single element $z$. In this case, the receptive field of $z$ on $\mathbf{Y}$ includes all the four elements of $\mathbf{Y}$, while the receptive field on the input includes all the nine input elements. Thus, when any element in a feature map needs a larger receptive field to detect input features over a broader area, we can build a deeper network.

Receptive fields derive their name from neurophysiology. A series of experiments on a range of animals using different stimuli (Hubel and Wiesel, 1959, Hubel and Wiesel, 1962, Hubel and Wiesel, 1968) explored the response of what is called the visual cortex on said stimuli. By and large they found that lower levels respond to edges and related shapes. Later on, Field (1987) illustrated this effect on natural images with, what can only be called, convolutional kernels. We reprint a key figure in Fig. 7.2.2 to illustrate the striking similarities.

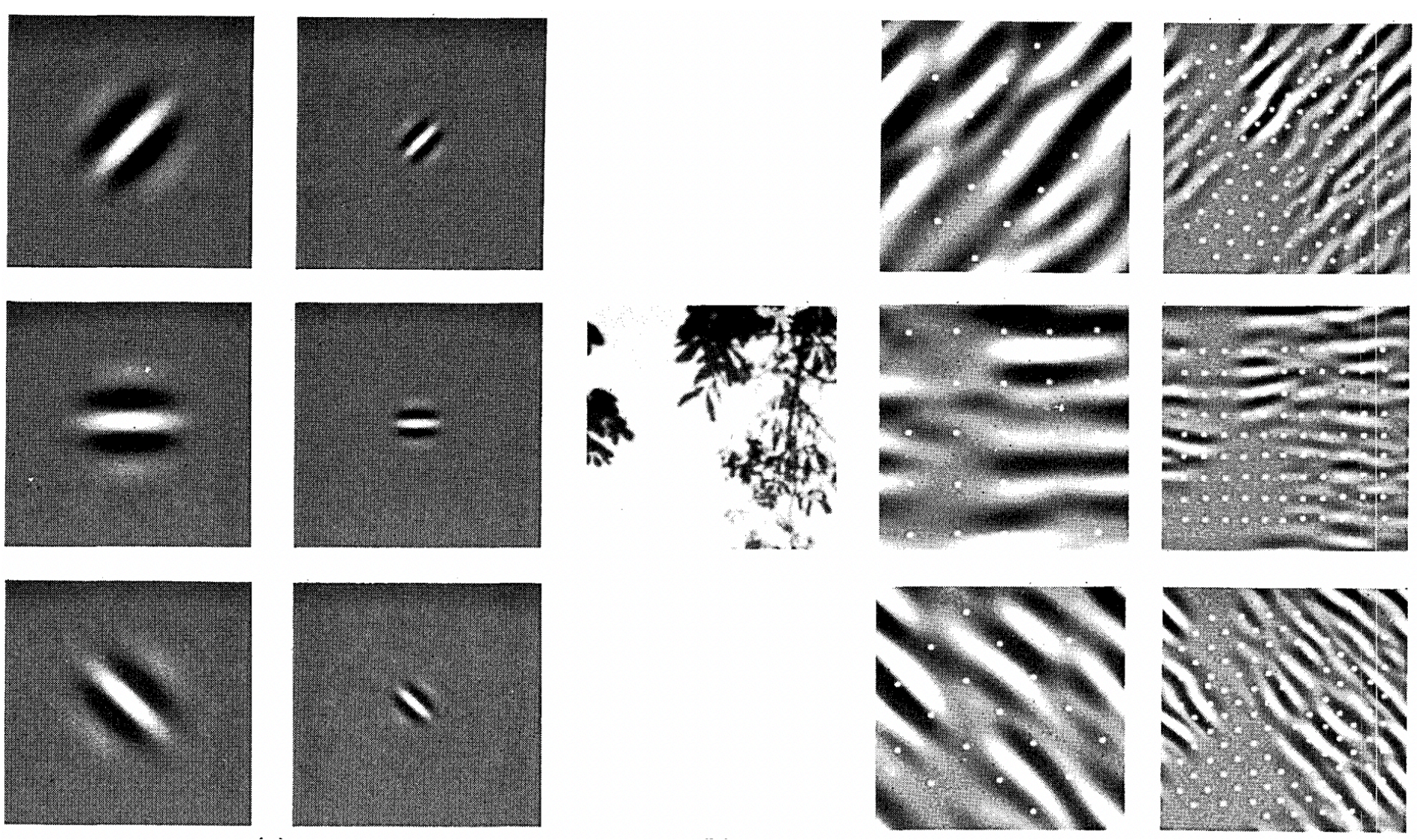

Fig. 7.2.2 Figure and caption taken from Field (1987): An example of coding with six different channels. (Left) Examples of the six types of sensor associated with each channel. (Right) Convolution of the image in (Middle) with the six sensors shown in (Left). The response of the individual sensors is determined by sampling these filtered images at a distance proportional to the size of the sensor (shown with dots). This diagram shows the response of only the even symmetric sensors.

As it turns out, this relation even holds for the features computed by deeper layers of networks trained on image classification tasks, as demonstrated in, for example, Kuzovkin *et al.* (2018). Suffice it to say, convolutions have proven to be an incredibly powerful tool for computer vision, both in biology and in code. As such, it is not surprising (in hindsight) that they heralded the recent success in deep learning.

## 7.2.7 Summary

The core computation required for a convolutional layer is a cross-correlation operation. We saw that a simple nested for-loop is all that is required to compute its value. If we have multiple input and multiple output channels, we are performing a matrix–matrix operation between channels. As can be seen, the computation is straightforward and, most importantly, highly *local*. This affords significant hardware optimization and many recent results in computer vision are only possible because of that. After all, it means that chip designers can invest in fast computation rather than memory when it comes to optimizing for convolutions. While this may not lead to optimal designs for other applications, it does open the door to ubiquitous and affordable computer vision.

In terms of convolutions themselves, they can be used for many purposes, for example detecting edges and lines, blurring images, or sharpening them. Most importantly, it is not necessary that the statistician (or engineer) invents suitable filters. Instead, we can simply *learn* them from data. This replaces feature engineering heuristics by evidence-based statistics. Lastly, and quite delightfully, these filters are not just advantageous for building deep networks but they also correspond to receptive fields and feature maps in the brain. This gives us confidence that we are on the right track.

## 7.2.8 Exercises

1. Construct an image `X` with diagonal edges.
    1. What happens if you apply the kernel `K` in this section to it?
    2. What happens if you transpose `X`?
    3. What happens if you transpose `K`?
2. Design some kernels manually.
    1. Given a directional vector $\mathbf{v} = (v_1, v_2)$, derive an edge-detection kernel that detects edges orthogonal to $\mathbf{v}$, i.e., edges in the direction $(v_2, -v_1)$.
    2. Derive a finite difference operator for the second derivative. What is the minimum size of the convolutional kernel associated with it? Which structures in images respond most strongly to it?
    3. How would you design a blur kernel? Why might you want to use such a kernel?
    4. What is the minimum size of a kernel to obtain a derivative of order $d$?
3. When you try to automatically find the gradient for the `Conv2D` class we created, what kind of error message do you see?
4. How do you represent a cross-correlation operation as a matrix multiplication by changing the input and kernel tensors?

Discussions[120].

# 7.3 Padding and Stride

Recall the example of a convolution in Fig. 7.2.1. The input had both a height and width of 3 and the convolution kernel had both a height and width of 2, yielding an output representation with dimension $2 \times 2$. Assuming that the input shape is $n_\mathrm{h} \times n_\mathrm{w}$ and the convolution kernel shape is $k_\mathrm{h} \times k_\mathrm{w}$, the output shape will be $(n_\mathrm{h} - k_\mathrm{h} + 1) \times (n_\mathrm{w} - k_\mathrm{w} + 1)$: we can only shift the convolution kernel so far until it runs out of pixels to apply the convolution to.

In the following we will explore a number of techniques, including padding and strided convolutions, that offer more control over the size of the output. As motivation, note that since kernels generally have width and height greater than 1, after applying many successive convolutions, we tend to wind up with outputs that are considerably smaller than our input. If we start with a $240 \times 240$ pixel image, ten layers of $5 \times 5$ convolutions reduce the image to $200 \times 200$ pixels, slicing off $30\%$ of the image and with it obliterating any interesting information on the boundaries of the original image. *Padding* is the most popular tool for handling this issue. In other cases, we may want to reduce the dimensionality drastically, e.g., if we find the original input resolution to be unwieldy. *Strided convolutions* are a popular technique that can help in these instances.

```
import torch
from torch import nn
```

## 7.3.1 Padding

As described above, one tricky issue when applying convolutional layers is that we tend to lose pixels on the perimeter of our image. Consider Fig. 7.3.1 that depicts the pixel utilization as a function of the convolution kernel size and the position within the image. The pixels in the corners are hardly used at all.

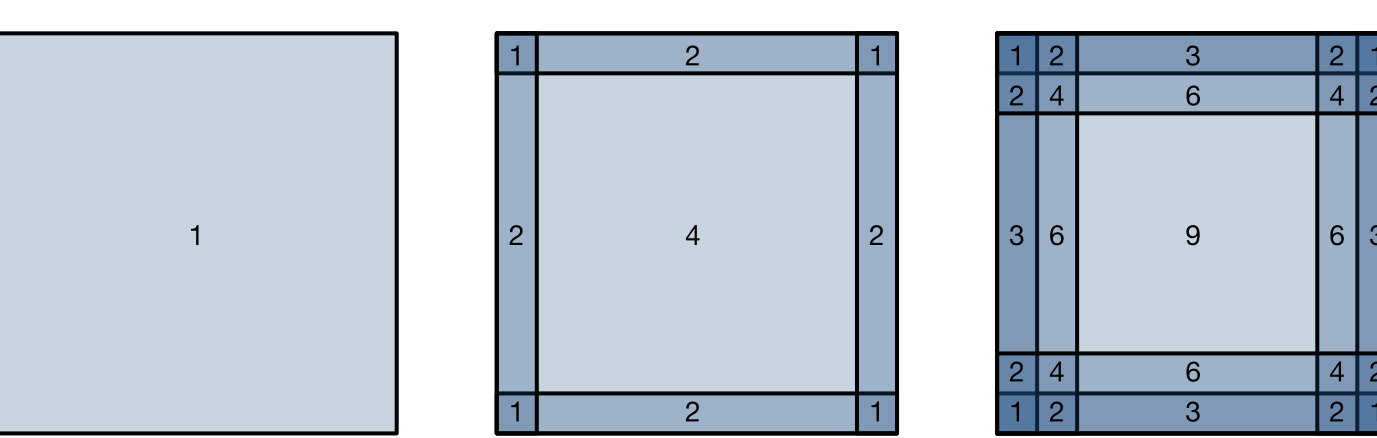

Fig. 7.3.1 Pixel utilization for convolutions of size $1 \times 1$, $2 \times 2$, and $3 \times 3$ respectively.

Since we typically use small kernels, for any given convolution we might only lose a few pixels but this can add up as we apply many successive convolutional layers. One straightforward solution to this problem is to add extra pixels of filler around the boundary of our input image, thus increasing the effective size of the image. Typically, we set the values of the extra pixels to zero. In Fig. 7.3.2, we pad a $3 \times 3$ input, increasing its size to $5 \times 5$. The corresponding output then increases to a $4 \times 4$ matrix. The shaded portions are the first output element as well as the input and kernel tensor elements used for the output computation: $0 \times 0 + 0 \times 1 + 0 \times 2 + 0 \times 3 = 0$.

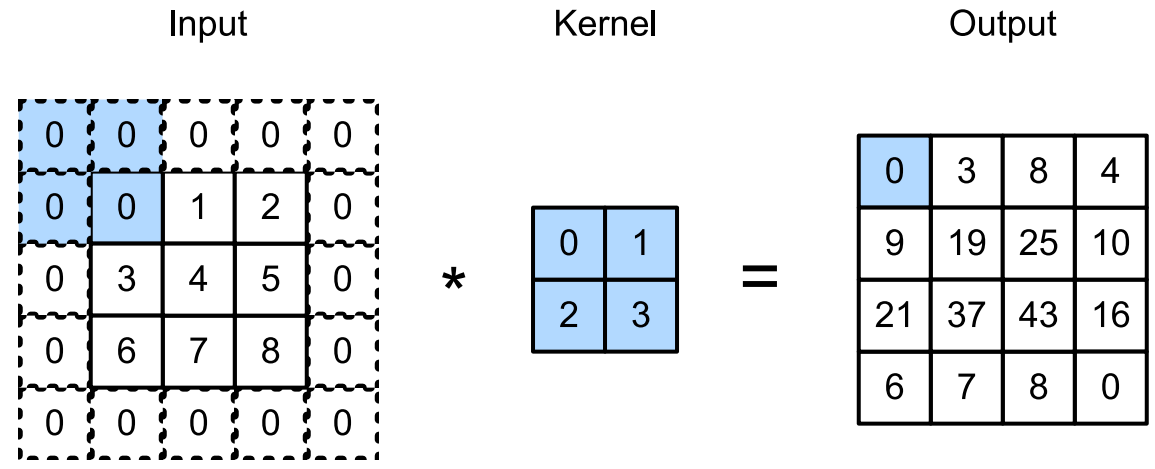

Fig. 7.3.2 Two-dimensional cross-correlation with padding.

In general, if we add a total of $p_\mathrm{h}$ rows of padding (roughly half on top and half on bottom) and a total of $p_\mathrm{w}$ columns of padding (roughly half on the left and half on the right), the output shape will be

$$(n_\mathrm{h}-k_\mathrm{h}+p_\mathrm{h}+1)\times(n_\mathrm{w}-k_\mathrm{w}+p_\mathrm{w}+1). \tag{7.3.1}$$

This means that the height and width of the output will increase by $p_\mathrm{h}$ and $p_\mathrm{w}$, respectively.

In many cases, we will want to set $p_\mathrm{h}=k_\mathrm{h}-1$ and $p_\mathrm{w}=k_\mathrm{w}-1$ to give the input and output the same height and width. This will make it easier to predict the output shape of each layer when constructing the network. Assuming that $k_\mathrm{h}$ is odd here, we will pad $p_\mathrm{h}/2$ rows on both sides of the height. If $k_\mathrm{h}$ is even, one possibility is to pad $\lceil p_\mathrm{h}/2\rceil$ rows on the top of the input and $\lfloor p_\mathrm{h}/2\rfloor$ rows on the bottom. We will pad both sides of the width in the same way.

CNNs commonly use convolution kernels with odd height and width values, such as 1, 3, 5, or 7. Choosing odd kernel sizes has the benefit that we can preserve the dimensionality while padding with the same number of rows on top and bottom, and the same number of columns on left and right.

Moreover, this practice of using odd kernels and padding to precisely preserve dimensionality offers a clerical benefit. For any two-dimensional tensor `X`, when the kernel's size is odd and the number of padding rows and columns on all sides are the same, thereby producing an output with the same height and width as the input, we know that the output `Y[i, j]` is calculated by cross-correlation of the input and convolution kernel with the window centered on `X[i, j]`.

In the following example, we create a two-dimensional convolutional layer with a height and width of 3 and apply 1 pixel of padding on all sides. Given an input with a height and width of 8, we find that the height and width of the output is also 8.

```
# We define a helper function to calculate convolutions. It initializes the
# convolutional layer weights and performs corresponding dimensionality
```

```
# elevations and reductions on the input and output
def comp_conv2d(conv2d, X):
    # (1, 1) indicates that batch size and the number of channels are both 1
    X = X.reshape((1, 1) + X.shape)
    Y = conv2d(X)
    # Strip the first two dimensions: examples and channels
    return Y.reshape(Y.shape[2:])

# 1 row and column is padded on either side, so a total of 2 rows or columns
# are added
conv2d = nn.LazyConv2d(1, kernel_size=3, padding=1)
X = torch.rand(size=(8, 8))
comp_conv2d(conv2d, X).shape
```

```
torch.Size([8, 8])
```

When the height and width of the convolution kernel are different, we can make the output and input have the same height and width by setting different padding numbers for height and width.

```
# We use a convolution kernel with height 5 and width 3. The padding on either
# side of the height and width are 2 and 1, respectively
conv2d = nn.LazyConv2d(1, kernel_size=(5, 3), padding=(2, 1))
comp_conv2d(conv2d, X).shape
```

```
torch.Size([8, 8])
```

### 7.3.2 Stride

When computing the cross-correlation, we start with the convolution window at the upper-left corner of the input tensor, and then slide it over all locations both down and to the right. In the previous examples, we defaulted to sliding one element at a time. However, sometimes, either for computational efficiency or because we wish to downsample, we move our window more than one element at a time, skipping the intermediate locations. This is particularly useful if the convolution kernel is large since it captures a large area of the underlying image.

We refer to the number of rows and columns traversed per slide as *stride*. So far, we have used strides of 1, both for height and width. Sometimes, we may want to use a larger stride. Fig. 7.3.3 shows a two-dimensional cross-correlation operation with a stride of 3 vertically and 2 horizontally. The shaded portions are the output elements as well as the input and kernel tensor elements used for the output computation: $0 \times 0 + 0 \times 1 + 1 \times 2 + 2 \times 3 = 8$, $0 \times 0 + 6 \times 1 + 0 \times 2 + 0 \times 3 = 6$. We can see that when the second element of the first column is generated, the convolution window slides down three rows. The convolution window slides two columns to the right when the second element of the first row is generated. When the convolution window continues to slide two columns to the right on the input, there is no output because the input element cannot fill the window (unless we add another column of padding).

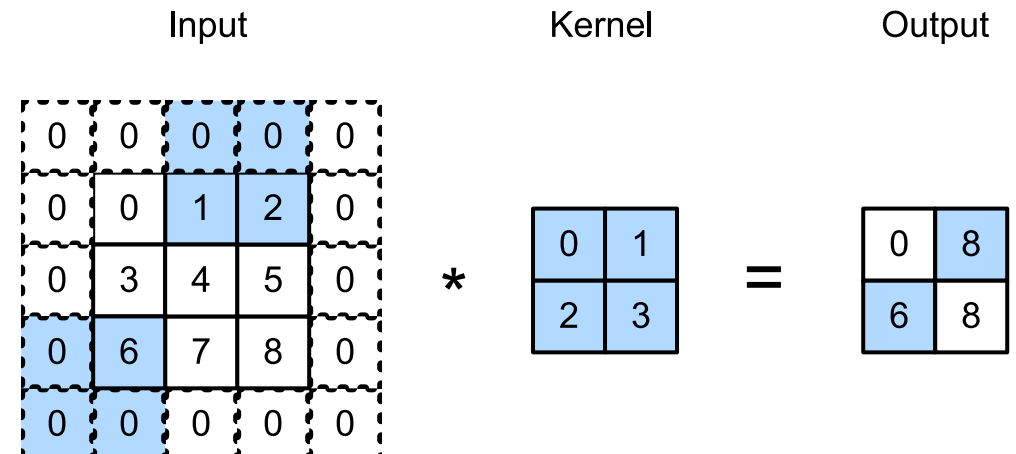

Fig. 7.3.3 Cross-correlation with strides of 3 and 2 for height and width, respectively.

In general, when the stride for the height is $s_\mathrm{h}$ and the stride for the width is $s_\mathrm{w}$, the output shape is

$$\lfloor(n_\mathrm{h}-k_\mathrm{h}+p_\mathrm{h}+s_\mathrm{h})/s_\mathrm{h}\rfloor \times \lfloor(n_\mathrm{w}-k_\mathrm{w}+p_\mathrm{w}+s_\mathrm{w})/s_\mathrm{w}\rfloor. \tag{7.3.2}$$

If we set $p_\mathrm{h}=k_\mathrm{h}-1$ and $p_\mathrm{w}=k_\mathrm{w}-1$, then the output shape can be simplified to $\lfloor(n_\mathrm{h}+s_\mathrm{h}-1)/s_\mathrm{h}\rfloor \times \lfloor(n_\mathrm{w}+s_\mathrm{w}-1)/s_\mathrm{w}\rfloor$. Going a step further, if the input height and width are divisible by the strides on the height and width, then the output shape will be $(n_\mathrm{h}/s_\mathrm{h}) \times (n_\mathrm{w}/s_\mathrm{w})$.

Below, we set the strides on both the height and width to 2, thus halving the input height and width.

```
conv2d = nn.LazyConv2d(1, kernel_size=3, padding=1, stride=2)
comp_conv2d(conv2d, X).shape
```

```
torch.Size([4, 4])
```

Let's look at a slightly more complicated example.

```
conv2d = nn.LazyConv2d(1, kernel_size=(3, 5), padding=(0, 1), stride=(3, 4))
comp_conv2d(conv2d, X).shape
```

```
torch.Size([2, 2])
```

### 7.3.3 Summary and Discussion

Padding can increase the height and width of the output. This is often used to give the output the same height and width as the input to avoid undesirable shrinkage of the output. Moreover, it ensures that all pixels are used equally frequently. Typically we pick symmetric padding on both sides of the input height and width. In this case we refer to $(p_\mathrm{h}, p_\mathrm{w})$ padding. Most commonly we set $p_\mathrm{h}=p_\mathrm{w}$, in which case we simply state that we choose padding $p$.

A similar convention applies to strides. When horizontal stride $s_\mathrm{h}$ and vertical stride $s_\mathrm{w}$ match,

we simply talk about stride $s$. The stride can reduce the resolution of the output, for example reducing the height and width of the output to only $1/n$ of the height and width of the input for $n > 1$. By default, the padding is 0 and the stride is 1.

So far all padding that we discussed simply extended images with zeros. This has significant computational benefit since it is trivial to accomplish. Moreover, operators can be engineered to take advantage of this padding implicitly without the need to allocate additional memory. At the same time, it allows CNNs to encode implicit position information within an image, simply by learning where the "whitespace" is. There are many alternatives to zero-padding. Alsallakh *et al.* (2020) provided an extensive overview of those (albeit without a clear case for when to use nonzero paddings unless artifacts occur).

### 7.3.4 Exercises

1. Given the final code example in this section with kernel size $(3, 5)$, padding $(0, 1)$, and stride $(3, 4)$, calculate the output shape to check if it is consistent with the experimental result.
2. For audio signals, what does a stride of 2 correspond to?
3. Implement mirror padding, i.e., padding where the border values are simply mirrored to extend tensors.
4. What are the computational benefits of a stride larger than 1?
5. What might be statistical benefits of a stride larger than 1?
6. How would you implement a stride of $\frac{1}{2}$? What does it correspond to? When would this be useful?

Discussions[121].

## 7.4 Multiple Input and Multiple Output Channels

While we described the multiple channels that comprise each image (e.g., color images have the standard RGB channels to indicate the amount of red, green and blue) and convolutional layers for multiple channels in Section 7.1.4, until now, we simplified all of our numerical examples by working with just a single input and a single output channel. This allowed us to think of our inputs, convolution kernels, and outputs each as two-dimensional tensors.

When we add channels into the mix, our inputs and hidden representations both become three-dimensional tensors. For example, each RGB input image has shape $3 \times h \times w$. We refer to this axis, with a size of 3, as the *channel* dimension. The notion of channels is as old as CNNs themselves: for instance LeNet-5 (LeCun *et al.*, 1995) uses them. In this section, we will take a deeper look at convolution kernels with multiple input and multiple output channels.

```
import torch
from d2l import torch as d2l
```

## 7.4.1 Multiple Input Channels

When the input data contains multiple channels, we need to construct a convolution kernel with the same number of input channels as the input data, so that it can perform cross-correlation with the input data. Assuming that the number of channels for the input data is $c_\mathrm{i}$, the number of input channels of the convolution kernel also needs to be $c_\mathrm{i}$. If our convolution kernel's window shape is $k_\mathrm{h} \times k_\mathrm{w}$, then, when $c_\mathrm{i} = 1$, we can think of our convolution kernel as just a two-dimensional tensor of shape $k_\mathrm{h} \times k_\mathrm{w}$.

However, when $c_\mathrm{i} > 1$, we need a kernel that contains a tensor of shape $k_\mathrm{h} \times k_\mathrm{w}$ for *every* input channel. Concatenating these $c_\mathrm{i}$ tensors together yields a convolution kernel of shape $c_\mathrm{i} \times k_\mathrm{h} \times k_\mathrm{w}$. Since the input and convolution kernel each have $c_\mathrm{i}$ channels, we can perform a cross-correlation operation on the two-dimensional tensor of the input and the two-dimensional tensor of the convolution kernel for each channel, adding the $c_\mathrm{i}$ results together (summing over the channels) to yield a two-dimensional tensor. This is the result of a two-dimensional cross-correlation between a multi-channel input and a multi-input-channel convolution kernel.

Fig. 7.4.1 provides an example of a two-dimensional cross-correlation with two input channels. The shaded portions are the first output element as well as the input and kernel tensor elements used for the output computation: $(1 \times 1 + 2 \times 2 + 4 \times 3 + 5 \times 4) + (0 \times 0 + 1 \times 1 + 3 \times 2 + 4 \times 3) = 56$.

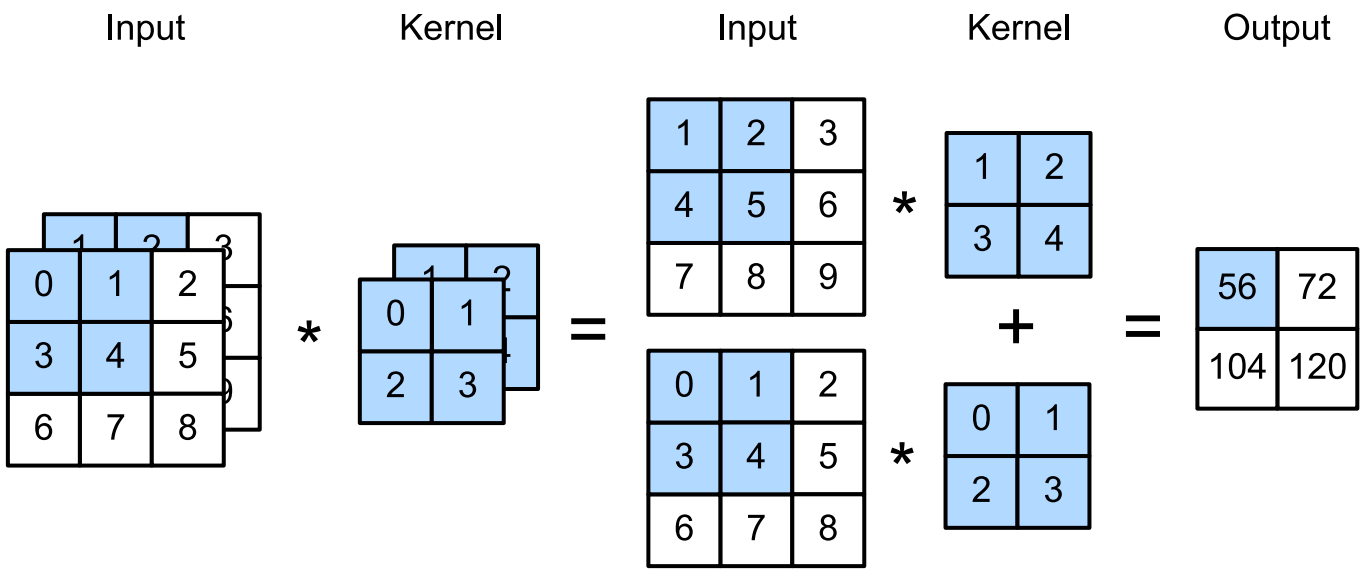

Fig. 7.4.1 Cross-correlation computation with two input channels.

To make sure we really understand what is going on here, we can implement cross-correlation operations with multiple input channels ourselves. Notice that all we are doing is performing a cross-correlation operation per channel and then adding up the results.

```
def corr2d_multi_in(X, K):
    # Iterate through the 0th dimension (channel) of K first, then add them up
    return sum(d2l.corr2d(x, k) for x, k in zip(X, K))
```

We can construct the input tensor X and the kernel tensor K corresponding to the values in Fig. 7.4.1 to validate the output of the cross-correlation operation.

```
X = torch.tensor([[[0.0, 1.0, 2.0], [3.0, 4.0, 5.0], [6.0, 7.0, 8.0]],
               [[1.0, 2.0, 3.0], [4.0, 5.0, 6.0], [7.0, 8.0, 9.0]]])
K = torch.tensor([[[0.0, 1.0], [2.0, 3.0]], [[1.0, 2.0], [3.0, 4.0]]])

corr2d_multi_in(X, K)
```

```
tensor([[ 56.,  72.],
        [104., 120.]])
```

### 7.4.2 Multiple Output Channels

Regardless of the number of input channels, so far we always ended up with one output channel. However, as we discussed in Section 7.1.4, it turns out to be essential to have multiple channels at each layer. In the most popular neural network architectures, we actually increase the channel dimension as we go deeper in the neural network, typically downsampling to trade off spatial resolution for greater *channel depth*. Intuitively, you could think of each channel as responding to a different set of features. The reality is a bit more complicated than this. A naive interpretation would suggest that representations are learned independently per pixel or per channel. Instead, channels are optimized to be jointly useful. This means that rather than mapping a single channel to an edge detector, it may simply mean that some direction in channel space corresponds to detecting edges.

Denote by $c_\textrm{i}$ and $c_\textrm{o}$ the number of input and output channels, respectively, and by $k_\textrm{h}$ and $k_\textrm{w}$ the height and width of the kernel. To get an output with multiple channels, we can create a kernel tensor of shape $c_\textrm{i}\times k_\textrm{h}\times k_\textrm{w}$ for *every* output channel. We concatenate them on the output channel dimension, so that the shape of the convolution kernel is $c_\textrm{o}\times c_\textrm{i}\times k_\textrm{h}\times k_\textrm{w}$. In cross-correlation operations, the result on each output channel is calculated from the convolution kernel corresponding to that output channel and takes input from all channels in the input tensor.

We implement a cross-correlation function to calculate the output of multiple channels as shown below.

```
def corr2d_multi_in_out(X, K):
    # Iterate through the 0th dimension of K, and each time, perform
    # cross-correlation operations with input X. All of the results are
    # stacked together
    return torch.stack([corr2d_multi_in(X, k) for k in K], 0)
```

We construct a trivial convolution kernel with three output channels by concatenating the kernel tensor for K with K+1 and K+2.

```
K = torch.stack((K, K + 1, K + 2), 0)
K.shape
```

```
torch.Size([3, 2, 2, 2])
```

Below, we perform cross-correlation operations on the input tensor X with the kernel tensor K. Now the output contains three channels. The result of the first channel is consistent with the result of the previous input tensor X and the multi-input channel, single-output channel kernel.

```
corr2d_multi_in_out(X, K)
```

```
tensor([[[ 56.,  72.],
         [104., 120.]],

        [[ 76., 100.],
         [148., 172.]],

        [[ 96., 128.],
         [192., 224.]]])
```

### 7.4.3 $1 \times 1$ Convolutional Layer

At first, a $1 \times 1$ convolution, i.e., $k_\textrm{h} = k_\textrm{w} = 1$, does not seem to make much sense. After all, a convolution correlates adjacent pixels. A $1 \times 1$ convolution obviously does not. Nonetheless, they are popular operations that are sometimes included in the designs of complex deep networks (Lin *et al.*, 2013, Szegedy *et al.*, 2017). Let's see in some detail what it actually does.

Because the minimum window is used, the $1 \times 1$ convolution loses the ability of larger convolutional layers to recognize patterns consisting of interactions among adjacent elements in the height and width dimensions. The only computation of the $1 \times 1$ convolution occurs on the channel dimension.

Fig. 7.4.2 shows the cross-correlation computation using the $1 \times 1$ convolution kernel with 3 input channels and 2 output channels. Note that the inputs and outputs have the same height and width. Each element in the output is derived from a linear combination of elements *at the same position* in the input image. You could think of the $1 \times 1$ convolutional layer as constituting a fully connected layer applied at every single pixel location to transform the $c_\textrm{i}$ corresponding input values into $c_\textrm{o}$ output values. Because this is still a convolutional layer, the weights are tied across pixel location. Thus the $1 \times 1$ convolutional layer requires $c_\textrm{o} \times c_\textrm{i}$ weights (plus the bias). Also note that convolutional layers are typically followed by nonlinearities. This ensures that $1 \times 1$ convolutions cannot simply be folded into other convolutions.

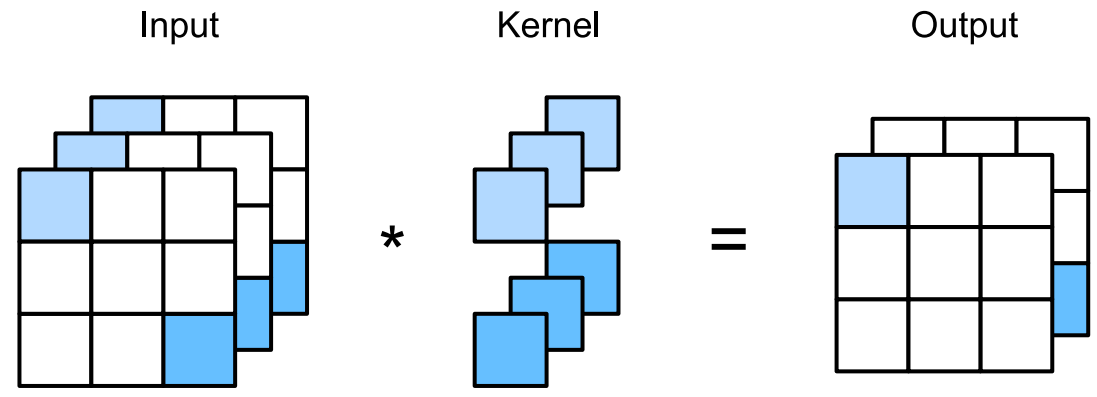

Fig. 7.4.2 The cross-correlation computation uses the $1 \times 1$ convolution kernel with three input channels and two output channels. The input and output have the same height and width.

Let's check whether this works in practice: we implement a $1 \times 1$ convolution using a fully connected layer. The only thing is that we need to make some adjustments to the data shape before and after the matrix multiplication.

```
def corr2d_multi_in_out_1x1(X, K):
    c_i, h, w = X.shape
    c_o = K.shape[0]
    X = X.reshape((c_i, h * w))
    K = K.reshape((c_o, c_i))
    # Matrix multiplication in the fully connected layer
    Y = torch.matmul(K, X)
    return Y.reshape((c_o, h, w))
```

When performing $1 \times 1$ convolutions, the above function is equivalent to the previously implemented cross-correlation function `corr2d_multi_in_out`. Let's check this with some sample data.

```
X = torch.normal(0, 1, (3, 3, 3))
K = torch.normal(0, 1, (2, 3, 1, 1))
Y1 = corr2d_multi_in_out_1x1(X, K)
Y2 = corr2d_multi_in_out(X, K)
assert float(torch.abs(Y1 - Y2).sum()) < 1e-6
```

## 7.4.4 Discussion

Channels allow us to combine the best of both worlds: MLPs that allow for significant nonlinearities and convolutions that allow for *localized* analysis of features. In particular, channels allow the CNN to reason with multiple features, such as edge and shape detectors at the same time. They also offer a practical trade-off between the drastic parameter reduction arising from translation invariance and locality, and the need for expressive and diverse models in computer vision.

Note, though, that this flexibility comes at a price. Given an image of size $(h \times w)$, the cost for computing a $k \times k$ convolution is $O(h \cdot w \cdot k^2)$. For $c_\textrm{i}$ and $c_\textrm{o}$ input and output channels respectively this increases to $O(h \cdot w \cdot k^2 \cdot c_\textrm{i} \cdot c_\textrm{o})$. For a $256 \times 256$ pixel image with a $5 \times 5$ kernel and $128$ input and output channels respectively this amounts to over 53 billion operations (we count multiplications and additions separately). Later on we will encounter effective strategies to cut down on the cost, e.g., by requiring the channel-wise operations to be block-diagonal, leading to architectures such as ResNeXt (Xie *et al.*, 2017).

### 7.4.5 Exercises

1. Assume that we have two convolution kernels of size $k_1$ and $k_2$, respectively (with no nonlinearity in between).
    1. Prove that the result of the operation can be expressed by a single convolution.
    2. What is the dimensionality of the equivalent single convolution?
    3. Is the converse true, i.e., can you always decompose a convolution into two smaller ones?
2. Assume an input of shape $c_\textrm{i}\times h\times w$ and a convolution kernel of shape $c_\textrm{o}\times c_\textrm{i}\times k_\textrm{h}\times k_\textrm{w}$, padding of $(p_\textrm{h}, p_\textrm{w})$, and stride of $(s_\textrm{h}, s_\textrm{w})$.
    1. What is the computational cost (multiplications and additions) for the forward propagation?
    2. What is the memory footprint?
    3. What is the memory footprint for the backward computation?
    4. What is the computational cost for the backpropagation?
3. By what factor does the number of calculations increase if we double both the number of input channels $c_\textrm{i}$ and the number of output channels $c_\textrm{o}$? What happens if we double the padding?
4. Are the variables `Y1` and `Y2` in the final example of this section exactly the same? Why?
5. Express convolutions as a matrix multiplication, even when the convolution window is not $1\times 1$.
6. Your task is to implement fast convolutions with a $k \times k$ kernel. One of the algorithm candidates is to scan horizontally across the source, reading a $k$-wide strip and computing the 1-wide output strip one value at a time. The alternative is to read a $k + \Delta$ wide strip and compute a $\Delta$-wide output strip. Why is the latter preferable? Is there a limit to how large you should choose $\Delta$?
7. Assume that we have a $c \times c$ matrix.
    1. How much faster is it to multiply with a block-diagonal matrix if the matrix is broken up into $b$ blocks?
    2. What is the downside of having $b$ blocks? How could you fix it, at least partly?

Discussions[122].

# 7.5 Pooling

In many cases our ultimate task asks some global question about the image, e.g., *does it contain a cat?* Consequently, the units of our final layer should be sensitive to the entire input. By gradually aggregating information, yielding coarser and coarser maps, we accomplish this goal of ultimately learning a global representation, while keeping all of the advantages of convolutional layers at the intermediate layers of processing. The deeper we go in the network, the larger the receptive field (relative to the input) to which each hidden node is sensitive. Reducing spatial resolution accelerates this process, since the convolution kernels cover a larger effective area.

Moreover, when detecting lower-level features, such as edges (as discussed in Section 7.2), we often want our representations to be somewhat invariant to translation. For instance, if we take the image `X` with a sharp delineation between black and white and shift the whole image by one pixel to the right, i.e., `Z[i, j] = X[i, j + 1]`, then the output for the new image `Z` might be vastly different. The edge will have shifted by one pixel. In reality, objects hardly ever occur exactly at the same place. In fact, even with a tripod and a stationary object, vibration of the camera due to the movement of the shutter might shift everything by a pixel or so (high-end cameras are loaded with special features to address this problem).

This section introduces *pooling layers*, which serve the dual purposes of mitigating the sensitivity of convolutional layers to location and of spatially downsampling representations.

```
import torch
from torch import nn
from d2l import torch as d2l
```

## 7.5.1 Maximum Pooling and Average Pooling

Like convolutional layers, *pooling* operators consist of a fixed-shape window that is slid over all regions in the input according to its stride, computing a single output for each location traversed by the fixed-shape window (sometimes known as the *pooling window*). However, unlike the cross-correlation computation of the inputs and kernels in the convolutional layer, the pooling layer contains no parameters (there is no *kernel*). Instead, pooling operators are deterministic, typically calculating either the maximum or the average value of the elements in the pooling window. These operations are called *maximum pooling* (*max-pooling* for short) and *average pooling*, respectively.

*Average pooling* is essentially as old as CNNs. The idea is akin to downsampling an image. Rather than just taking the value of every second (or third) pixel for the lower resolution image, we can average over adjacent pixels to obtain an image with better signal-to-noise ratio since we are combining the information from multiple adjacent pixels. *Max-pooling* was introduced in Riesenhuber and Poggio (1999) in the context of cognitive neuroscience to describe how information aggregation might be aggregated hierarchically for the purpose of object recognition; there already was an earlier version in speech recognition (Yamaguchi *et al.*, 1990). In almost all cases, max-pooling, as it is also referred to, is preferable to average pooling.

In both cases, as with the cross-correlation operator, we can think of the pooling window as starting from the upper-left of the input tensor and sliding across it from left to right and top to bottom. At each location that the pooling window hits, it computes the maximum or average value of the input subtensor in the window, depending on whether max or average pooling is employed.

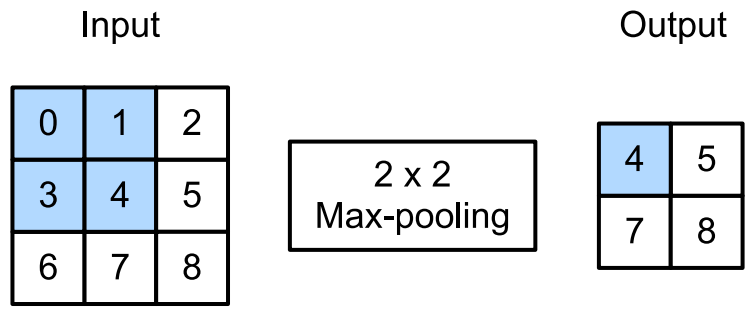

Fig. 7.5.1 Max-pooling with a pooling window shape of $2 \times 2$. The shaded portions are the first output element as well as the input tensor elements used for the output computation: $\max(0, 1, 3, 4) = 4$.

The output tensor in Fig. 7.5.1 has a height of 2 and a width of 2. The four elements are derived from the maximum value in each pooling window:

$$\begin{aligned}\max(0, 1, 3, 4) &= 4,\\ \max(1, 2, 4, 5) &= 5,\\ \max(3, 4, 6, 7) &= 7,\\ \max(4, 5, 7, 8) &= 8.\end{aligned} \tag{7.5.1}$$

More generally, we can define a $p \times q$ pooling layer by aggregating over a region of said size. Returning to the problem of edge detection, we use the output of the convolutional layer as input for $2 \times 2$ max-pooling. Denote by `X` the input of the convolutional layer input and `Y` the pooling layer output. Regardless of whether or not the values of `X[i, j]`, `X[i, j + 1]`, `X[i+1, j]` and `X[i+1, j + 1]` are different, the pooling layer always outputs `Y[i, j] = 1`. That is to say, using the $2 \times 2$ max-pooling layer, we can still detect if the pattern recognized by the convolutional layer moves no more than one element in height or width.

In the code below, we implement the forward propagation of the pooling layer in the `pool2d` function. This function is similar to the `corr2d` function in Section 7.2. However, no kernel is needed, computing the output as either the maximum or the average of each region in the input.

```
def pool2d(X, pool_size, mode='max'):
    p_h, p_w = pool_size
    Y = torch.zeros((X.shape[0] - p_h + 1, X.shape[1] - p_w + 1))
    for i in range(Y.shape[0]):
        for j in range(Y.shape[1]):
            if mode == 'max':
```

```
                Y[i, j] = X[i: i + p_h, j: j + p_w].max()
            elif mode == 'avg':
                Y[i, j] = X[i: i + p_h, j: j + p_w].mean()
    return Y
```

We can construct the input tensor `X` in Fig. 7.5.1 to validate the output of the two-dimensional max-pooling layer.

```
X = torch.tensor([[0.0, 1.0, 2.0], [3.0, 4.0, 5.0], [6.0, 7.0, 8.0]])
pool2d(X, (2, 2))
```

```
tensor([[4., 5.],
        [7., 8.]])
```

Also, we can experiment with the average pooling layer.

```
pool2d(X, (2, 2), 'avg')
```

```
tensor([[2., 3.],
        [5., 6.]])
```

## 7.5.2 Padding and Stride

As with convolutional layers, pooling layers change the output shape. And as before, we can adjust the operation to achieve a desired output shape by padding the input and adjusting the stride. We can demonstrate the use of padding and strides in pooling layers via the built-in two-dimensional max-pooling layer from the deep learning framework. We first construct an input tensor `X` whose shape has four dimensions, where the number of examples (batch size) and number of channels are both 1.

```
X = torch.arange(16, dtype=torch.float32).reshape((1, 1, 4, 4))
X
```

```
tensor([[[[ 0.,  1.,  2.,  3.],
          [ 4.,  5.,  6.,  7.],
          [ 8.,  9., 10., 11.],
          [12., 13., 14., 15.]]]])
```

Since pooling aggregates information from an area, deep learning frameworks default to matching pooling window sizes and stride. For instance, if we use a pooling window of shape `(3, 3)` we get a stride shape of `(3, 3)` by default.

```
pool2d = nn.MaxPool2d(3)
# Pooling has no model parameters, hence it needs no initialization
pool2d(X)
```

```
tensor([[[[10.]]]])
```

Needless to say, the stride and padding can be manually specified to override framework defaults if required.

```
pool2d = nn.MaxPool2d(3, padding=1, stride=2)
pool2d(X)
```

```
tensor([[[[ 5.,  7.],
          [13., 15.]]]])
```

Of course, we can specify an arbitrary rectangular pooling window with arbitrary height and width respectively, as the example below shows.

```
pool2d = nn.MaxPool2d((2, 3), stride=(2, 3), padding=(0, 1))
pool2d(X)
```

```
tensor([[[[ 5.,  7.],
          [13., 15.]]]])
```

### 7.5.3 Multiple Channels

When processing multi-channel input data, the pooling layer pools each input channel separately, rather than summing the inputs up over channels as in a convolutional layer. This means that the number of output channels for the pooling layer is the same as the number of input channels. Below, we will concatenate tensors `X` and `X + 1` on the channel dimension to construct an input with two channels.

```
X = torch.cat((X, X + 1), 1)
X
```

```
tensor([[[[ 0.,  1.,  2.,  3.],
          [ 4.,  5.,  6.,  7.],
          [ 8.,  9., 10., 11.],
          [12., 13., 14., 15.]],

         [[ 1.,  2.,  3.,  4.],
          [ 5.,  6.,  7.,  8.],
          [ 9., 10., 11., 12.],
          [13., 14., 15., 16.]]]])
```

As we can see, the number of output channels is still two after pooling.

```
pool2d = nn.MaxPool2d(3, padding=1, stride=2)
pool2d(X)
```

```
tensor([[[[ 5.,  7.],
          [13., 15.]],

         [[ 6.,  8.],
          [14., 16.]]]])
```

### 7.5.4 Summary

Pooling is an exceedingly simple operation. It does exactly what its name indicates, aggregate results over a window of values. All convolution semantics, such as strides and padding apply in the same way as they did previously. Note that pooling is indifferent to channels, i.e., it leaves the number of channels unchanged and it applies to each channel separately. Lastly, of the two popular pooling choices, max-pooling is preferable to average pooling, as it confers some degree of invariance to output. A popular choice is to pick a pooling window size of $2 \times 2$ to quarter the spatial resolution of output.

Note that there are many more ways of reducing resolution beyond pooling. For instance, in stochastic pooling (Zeiler and Fergus, 2013) and fractional max-pooling (Graham, 2014) aggregation is combined with randomization. This can slightly improve the accuracy in some cases. Lastly, as we will see later with the attention mechanism, there are more refined ways of aggregating over outputs, e.g., by using the alignment between a query and representation vectors.

### 7.5.5 Exercises

1. Implement average pooling through a convolution.
2. Prove that max-pooling cannot be implemented through a convolution alone.
3. Max-pooling can be accomplished using ReLU operations, i.e., $\textrm{ReLU}(x) = \max(0, x)$.
    1. Express $\max(a, b)$ by using only ReLU operations.
    2. Use this to implement max-pooling by means of convolutions and ReLU layers.
    3. How many channels and layers do you need for a $2 \times 2$ convolution? How many for a $3 \times 3$ convolution?
4. What is the computational cost of the pooling layer? Assume that the input to the pooling layer is of size $c \times h \times w$, the pooling window has a shape of $p_\textrm{h} \times p_\textrm{w}$ with a padding of $(p_\textrm{h}, p_\textrm{w})$ and a stride of $(s_\textrm{h}, s_\textrm{w})$.
5. Why do you expect max-pooling and average pooling to work differently?

6. Do we need a separate minimum pooling layer? Can you replace it with another operation?
7. We could use the softmax operation for pooling. Why might it not be so popular?

Discussions[123].

# 7.6 Convolutional Neural Networks (LeNet)

We now have all the ingredients required to assemble a fully-functional CNN. In our earlier encounter with image data, we applied a linear model with softmax regression (Section 4.4) and an MLP (Section 5.2) to pictures of clothing in the Fashion-MNIST dataset. To make such data amenable we first flattened each image from a $28 \times 28$ matrix into a fixed-length $784$-dimensional vector, and thereafter processed them in fully connected layers. Now that we have a handle on convolutional layers, we can retain the spatial structure in our images. As an additional benefit of replacing fully connected layers with convolutional layers, we will enjoy more parsimonious models that require far fewer parameters.

In this section, we will introduce *LeNet*, among the first published CNNs to capture wide attention for its performance on computer vision tasks. The model was introduced by (and named for) Yann LeCun, then a researcher at AT&T Bell Labs, for the purpose of recognizing handwritten digits in images (LeCun *et al.*, 1998). This work represented the culmination of a decade of research developing the technology; LeCun's team published the first study to successfully train CNNs via backpropagation (LeCun *et al.*, 1989).

At the time LeNet achieved outstanding results matching the performance of support vector machines, then a dominant approach in supervised learning, achieving an error rate of less than 1% per digit. LeNet was eventually adapted to recognize digits for processing deposits in ATM machines. To this day, some ATMs still run the code that Yann LeCun and his colleague Leon Bottou wrote in the 1990s!

```
import torch
from torch import nn
from d2l import torch as d2l
```

## 7.6.1 LeNet

At a high level, LeNet (LeNet-5) consists of two parts: (i) a convolutional encoder consisting of two convolutional layers; and (ii) a dense block consisting of three fully connected layers. The architecture is summarized in Fig. 7.6.1.

The basic units in each convolutional block are a convolutional layer, a sigmoid activation function, and a subsequent average pooling operation. Note that while ReLUs and max-pooling work better, they had not yet been discovered. Each convolutional layer uses a $5 \times 5$ kernel and a sigmoid activation function. These layers map spatially arranged inputs to a number of two-dimensional feature maps, typically increasing the number of channels. The first convolutional layer has 6 output channels, while the second has 16. Each $2 \times 2$ pooling operation (stride 2) reduces dimensionality by a factor of 4 via spatial downsampling. The convolutional block emits an output with shape given by (batch size, number of channel, height, width).

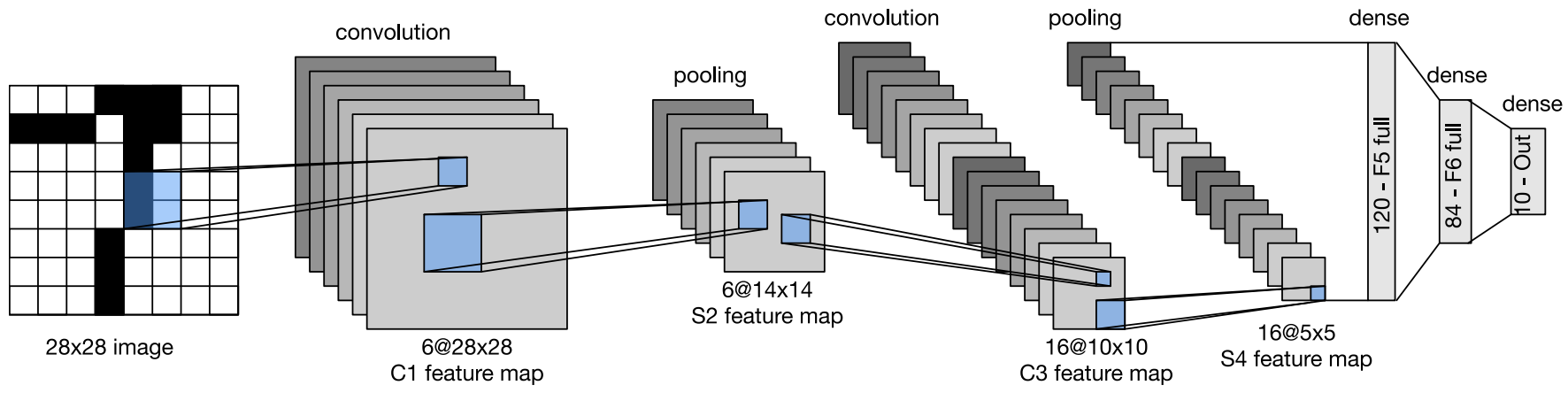

Fig. 7.6.1 Data flow in LeNet. The input is a handwritten digit, the output is a probability over 10 possible outcomes.

In order to pass output from the convolutional block to the dense block, we must flatten each example in the minibatch. In other words, we take this four-dimensional input and transform it into the two-dimensional input expected by fully connected layers: as a reminder, the two-dimensional representation that we desire uses the first dimension to index examples in the minibatch and the second to give the flat vector representation of each example. LeNet's dense block has three fully connected layers, with 120, 84, and 10 outputs, respectively. Because we are still performing classification, the 10-dimensional output layer corresponds to the number of possible output classes.

While getting to the point where you truly understand what is going on inside LeNet may have taken a bit of work, we hope that the following code snippet will convince you that implementing such models with modern deep learning frameworks is remarkably simple. We need only to instantiate a `Sequential` block and chain together the appropriate layers, using Xavier initialization as introduced in Section 5.4.2.

```
def init_cnn(module):  #@save
    """Initialize weights for CNNs."""
    if type(module) == nn.Linear or type(module) == nn.Conv2d:
        nn.init.xavier_uniform_(module.weight)
```

```
class LeNet(d2l.Classifier):  #@save
    """The LeNet-5 model."""
    def __init__(self, lr=0.1, num_classes=10):
        super().__init__()
        self.save_hyperparameters()
```

```
        self.net = nn.Sequential(
            nn.LazyConv2d(6, kernel_size=5, padding=2), nn.Sigmoid(),
            nn.AvgPool2d(kernel_size=2, stride=2),
            nn.LazyConv2d(16, kernel_size=5), nn.Sigmoid(),
            nn.AvgPool2d(kernel_size=2, stride=2),
            nn.Flatten(),
            nn.LazyLinear(120), nn.Sigmoid(),
            nn.LazyLinear(84), nn.Sigmoid(),
            nn.LazyLinear(num_classes))
```

We have taken some liberty in the reproduction of LeNet insofar as we have replaced the Gaussian activation layer by a softmax layer. This greatly simplifies the implementation, not least due to the fact that the Gaussian decoder is rarely used nowadays. Other than that, this network matches the original LeNet-5 architecture.

Let's see what happens inside the network. By passing a single-channel (black and white) $28 \times 28$ image through the network and printing the output shape at each layer, we can inspect the model to ensure that its operations line up with what we expect from Fig. 7.6.2.

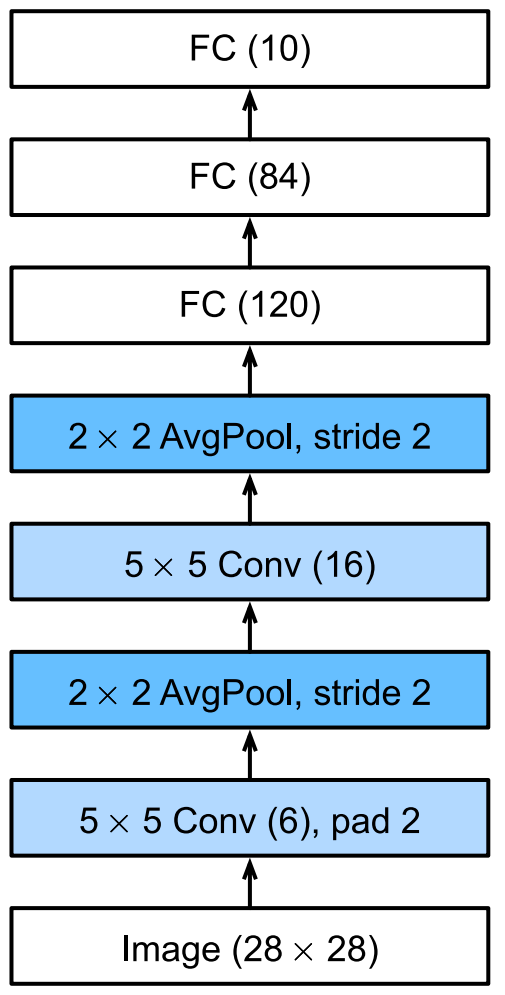

Fig. 7.6.2 Compressed notation for LeNet-5.

```
@d2l.add_to_class(d2l.Classifier)  #@save
def layer_summary(self, X_shape):
    X = torch.randn(*X_shape)
    for layer in self.net:
        X = layer(X)
        print(layer.__class__.__name__, 'output shape:\t', X.shape)

model = LeNet()
model.layer_summary((1, 1, 28, 28))
```

```
Conv2d output shape:         torch.Size([1, 6, 28, 28])
Sigmoid output shape:        torch.Size([1, 6, 28, 28])
AvgPool2d output shape:      torch.Size([1, 6, 14, 14])
Conv2d output shape:         torch.Size([1, 16, 10, 10])
Sigmoid output shape:        torch.Size([1, 16, 10, 10])
AvgPool2d output shape:      torch.Size([1, 16, 5, 5])
Flatten output shape:        torch.Size([1, 400])
Linear output shape:         torch.Size([1, 120])
Sigmoid output shape:        torch.Size([1, 120])
Linear output shape:         torch.Size([1, 84])
Sigmoid output shape:        torch.Size([1, 84])
Linear output shape:         torch.Size([1, 10])
```

Note that the height and width of the representation at each layer throughout the convolutional block is reduced (compared with the previous layer). The first convolutional layer uses two pixels of padding to compensate for the reduction in height and width that would otherwise result from using a $5 \times 5$ kernel. As an aside, the image size of $28 \times 28$ pixels in the original MNIST OCR dataset is a result of *trimming* two pixel rows (and columns) from the original scans that measured $32 \times 32$ pixels. This was done primarily to save space (a 30% reduction) at a time when megabytes mattered.

In contrast, the second convolutional layer forgoes padding, and thus the height and width are both reduced by four pixels. As we go up the stack of layers, the number of channels increases layer-over-layer from 1 in the input to 6 after the first convolutional layer and 16 after the second convolutional layer. However, each pooling layer halves the height and width. Finally, each fully connected layer reduces dimensionality, finally emitting an output whose dimension matches the number of classes.

## 7.6.2 Training

Now that we have implemented the model, let's run an experiment to see how the LeNet-5 model fares on Fashion-MNIST.

While CNNs have fewer parameters, they can still be more expensive to compute than similarly deep MLPs because each parameter participates in many more multiplications. If you have access to a GPU, this might be a good time to put it into action to speed up training. Note that the `d2l.Trainer` class takes care of all details. By default, it initializes the model parameters on the available devices. Just as with MLPs, our loss function is cross-entropy, and we minimize it via minibatch stochastic gradient descent.

```
trainer = d2l.Trainer(max_epochs=10, num_gpus=1)
data = d2l.FashionMNIST(batch_size=128)
model = LeNet(lr=0.1)
model.apply_init([next(iter(data.get_dataloader(True)))[0]], init_cnn)
trainer.fit(model, data)
```

## 7.6.3 Summary

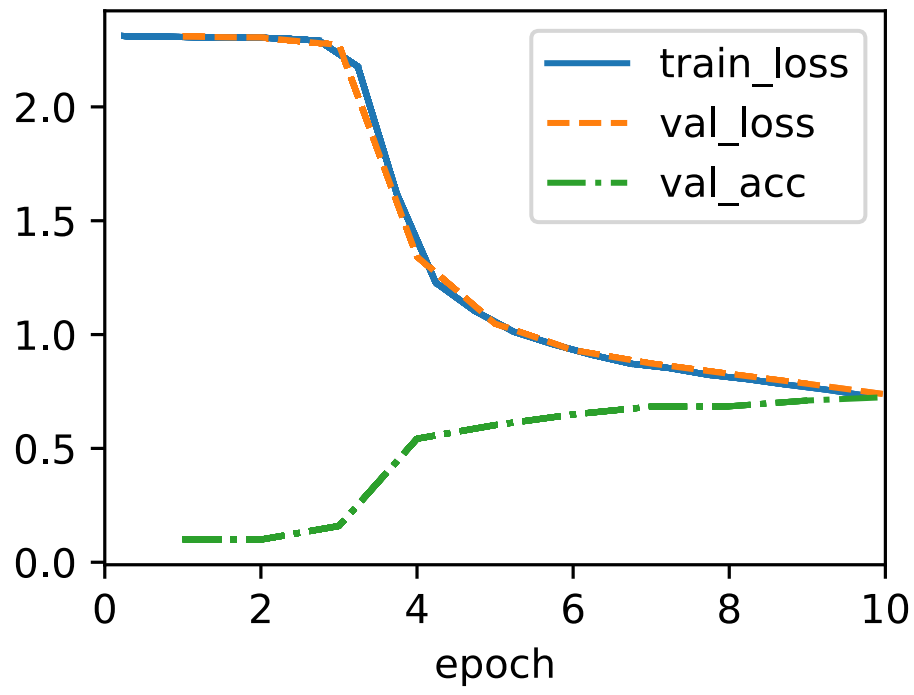

We have made significant progress in this chapter. We moved from the MLPs of the 1980s to the CNNs of the 1990s and early 2000s. The architectures proposed, e.g., in the form of LeNet-5 remain meaningful, even to this day. It is worth comparing the error rates on Fashion-MNIST achievable with LeNet-5 both to the very best possible with MLPs (Section 5.2) and those with significantly more advanced architectures such as ResNet (Section 8.6). LeNet is much more similar to the latter than to the former. One of the primary differences, as we shall see, is that greater amounts of computation enabled significantly more complex architectures.

A second difference is the relative ease with which we were able to implement LeNet. What used to be an engineering challenge worth months of C++ and assembly code, engineering to improve SN, an early Lisp-based deep learning tool (Bottou and Le Cun, 1988), and finally experimentation with models can now be accomplished in minutes. It is this incredible productivity boost that has democratized deep learning model development tremendously. In the next chapter we will journey down this rabbit to hole to see where it takes us.

### 7.6.4 Exercises

1. Let's modernize LeNet. Implement and test the following changes:
    1. Replace average pooling with max-pooling.
    2. Replace the softmax layer with ReLU.
2. Try to change the size of the LeNet style network to improve its accuracy in addition to max-pooling and ReLU.
    1. Adjust the convolution window size.
    2. Adjust the number of output channels.
    3. Adjust the number of convolution layers.
    4. Adjust the number of fully connected layers.

5. Adjust the learning rates and other training details (e.g., initialization and number of epochs).

3. Try out the improved network on the original MNIST dataset.

4. Display the activations of the first and second layer of LeNet for different inputs (e.g., sweaters and coats).

5. What happens to the activations when you feed significantly different images into the network (e.g., cats, cars, or even random noise)?

Discussions[124].

# 8 Modern Convolutional Neural Networks

Now that we understand the basics of wiring together CNNs, let's take a tour of modern CNN architectures. This tour is, by necessity, incomplete, thanks to the plethora of exciting new designs being added. Their importance derives from the fact that not only can they be used directly for vision tasks, but they also serve as basic feature generators for more advanced tasks such as tracking (Zhang *et al.*, 2021), segmentation (Long *et al.*, 2015), object detection (Redmon and Farhadi, 2018), or style transformation (Gatys *et al.*, 2016). In this chapter, most sections correspond to a significant CNN architecture that was at some point (or currently) the base model upon which many research projects and deployed systems were built. Each of these networks was briefly a dominant architecture and many were winners or runners-up in the ImageNet competition[125] which has served as a barometer of progress on supervised learning in computer vision since 2010. It is only recently that Transformers have begun to displace CNNs, starting with Dosovitskiy *et al.* (2021) and followed by the Swin Transformer (Liu *et al.*, 2021). We will cover this development later in Chapter 11.

While the idea of *deep* neural networks is quite simple (stack together a bunch of layers), performance can vary wildly across architectures and hyperparameter choices. The neural networks described in this chapter are the product of intuition, a few mathematical insights, and a lot of trial and error. We present these models in chronological order, partly to convey a sense of the history so that you can form your own intuitions about where the field is heading and perhaps develop your own architectures. For instance, batch normalization and residual connections described in this chapter have offered two popular ideas for training and designing deep models, both of which have since also been applied to architectures beyond computer vision.

We begin our tour of modern CNNs with AlexNet (Krizhevsky *et al.*, 2012), the first large-scale network deployed to beat conventional computer vision methods on a large-scale vision challenge; the VGG network (Simonyan and Zisserman, 2014), which makes use of a number of repeating blocks of elements; the network in network (NiN) that convolves whole neural networks patch-wise over inputs (Lin *et al.*, 2013); GoogLeNet that uses networks with multi-branch convolutions (Szegedy *et al.*, 2015); the residual network (ResNet) (He *et al.*, 2016), which remains one of the most popular off-the-shelf architectures in computer vision; ResNeXt blocks (Xie *et al.*, 2017) for sparser connections; and DenseNet (Huang *et al.*, 2017) for a generalization of the residual architecture. Over time many special optimizations for efficient networks have been developed, such as coordinate shifts (ShiftNet) (Wu *et al.*, 2018). This culminated in the automatic search for efficient architectures such as MobileNet v3 (Howard *et al.*, 2019). It also includes the semi-automatic design exploration of Radosavovic *et al.* (2020) that led to the RegNetX/Y which we will discuss later in this chapter. The work is instructive insofar as it offers a path for marrying brute force computation with the ingenuity of an experimenter in the search for efficient design spaces. Of note is also the work of Liu *et al.* (2022) as it shows that training techniques (e.g., optimizers, data augmentation, and regularization) play a pivotal role in improving accuracy. It also shows that long-held assumptions, such as the size of a convolution window, may need to be revisited, given the increase in computation and data. We will cover this and many more questions in due course throughout this chapter.

## 8.1 Deep Convolutional Neural Networks (AlexNet)

Although CNNs were well known in the computer vision and machine learning communities following the introduction of LeNet (LeCun *et al.*, 1995), they did not immediately dominate the field. Although LeNet achieved good results on early small datasets, the performance and feasibility of training CNNs on larger, more realistic datasets had yet to be established. In fact, for much of the intervening time between the early 1990s and the watershed results of 2012 (Krizhevsky *et al.*, 2012), neural networks were often surpassed by other machine learning methods, such as kernel methods (Schölkopf and Smola, 2002), ensemble methods (Freund and Schapire, 1996), and structured estimation (Taskar *et al.*, 2004).

For computer vision, this comparison is perhaps not entirely accurate. That is, although the inputs to convolutional networks consist of raw or lightly-processed (e.g., by centering) pixel values, practitioners would never feed raw pixels into traditional models. Instead, typical computer vision pipelines consisted of manually engineering feature extraction pipelines, such as SIFT (Lowe, 2004), SURF (Bay *et al.*, 2006), and bags of visual words (Sivic and Zisserman, 2003). Rather than *learning* the features, the features were *crafted*. Most of the progress came from having more clever ideas for feature extraction on the one hand and deep insight into geometry (Hartley and Zisserman, 2000) on the other. The learning algorithm was often considered an afterthought.

Although some neural network accelerators were available in the 1990s, they were not yet sufficiently powerful to make deep multichannel, multilayer CNNs with a large number of parameters. For instance, NVIDIA's GeForce 256 from 1999 was able to process at most 480 million floating-point operations, such as additions and multiplications, per second (MFLOPS), without any meaningful programming framework for operations beyond games. Today's accelerators are able to perform in excess of 1000 TFLOPs per device. Moreover, datasets were still relatively small: OCR on 60,000 low-resolution $28 \times 28$ pixel images was considered a highly challenging task. Added to these obstacles, key tricks for training neural networks including parameter initialization heuristics (Glorot and Bengio, 2010), clever variants of stochastic gradient descent (Kingma and Ba, 2014), non-squashing activation functions (Nair and Hinton, 2010), and effective regularization techniques (Srivastava *et al.*, 2014) were still missing.

Thus, rather than training *end-to-end* (pixel to classification) systems, classical pipelines looked more like this:

1. Obtain an interesting dataset. In the early days, these datasets required expensive sensors. For instance, the Apple QuickTake 100[126] of 1994 sported a whopping 0.3 megapixel (VGA) resolution, capable of storing up to 8 images, all for the price of $1000.

126

2. Preprocess the dataset with hand-crafted features based on some knowledge of optics, geometry, other analytic tools, and occasionally on the serendipitous discoveries by lucky graduate students.

3. Feed the data through a standard set of feature extractors such as the SIFT (scale-invariant feature transform) (Lowe, 2004), the SURF (speeded up robust features) (Bay *et al.*, 2006), or any number of other hand-tuned pipelines. OpenCV still provides SIFT extractors to this day!

4. Dump the resulting representations into your favorite classifier, likely a linear model or kernel method, to train a classifier.

If you spoke to machine learning researchers, they would reply that machine learning was both important and beautiful. Elegant theories proved the properties of various classifiers (Boucheron *et al.*, 2005) and convex optimization (Boyd and Vandenberghe, 2004) had become the mainstay for obtaining them. The field of machine learning was thriving, rigorous, and eminently useful. However, if you spoke to a computer vision researcher, you would hear a very different story. The dirty truth of image recognition, they would tell you, is that features, geometry (Hartley and Zisserman, 2000, Hartley and Kahl, 2009), and engineering, rather than novel learning algorithms, drove progress. Computer vision researchers justifiably believed that a slightly bigger or cleaner dataset or a slightly improved feature-extraction pipeline mattered far more to the final accuracy than any learning algorithm.

```
import torch
from torch import nn
from d2l import torch as d2l
```

### 8.1.1 Representation Learning

Another way to cast the state of affairs is that the most important part of the pipeline was the representation. And up until 2012 the representation was calculated mostly mechanically. In fact, engineering a new set of feature functions, improving results, and writing up the method all featured prominently in papers. SIFT (Lowe, 2004), SURF (Bay *et al.*, 2006), HOG (histograms of oriented gradient) (Dalal and Triggs, 2005), bags of visual words (Sivic and Zisserman, 2003), and similar feature extractors ruled the roost.

Another group of researchers, including Yann LeCun, Geoff Hinton, Yoshua Bengio, Andrew Ng, Shun-ichi Amari, and Juergen Schmidhuber, had different plans. They believed that features themselves ought to be learned. Moreover, they believed that to be reasonably complex, the features ought to be hierarchically composed with multiple jointly learned layers, each with learnable parameters. In the case of an image, the lowest layers might come to detect edges, colors, and textures, by analogy with how the visual system in animals processes its input. In particular, the automatic design of visual features such as those obtained by sparse coding (Olshausen and Field, 1996) remained an open challenge until the advent of modern CNNs. It was not until Dean *et al.* (2012), Le (2013) that the idea of generating features from image data automatically gained significant traction.

The first modern CNN (Krizhevsky *et al.*, 2012), named *AlexNet* after one of its inventors, Alex Krizhevsky, is largely an evolutionary improvement over LeNet. It achieved excellent performance in the 2012 ImageNet challenge.

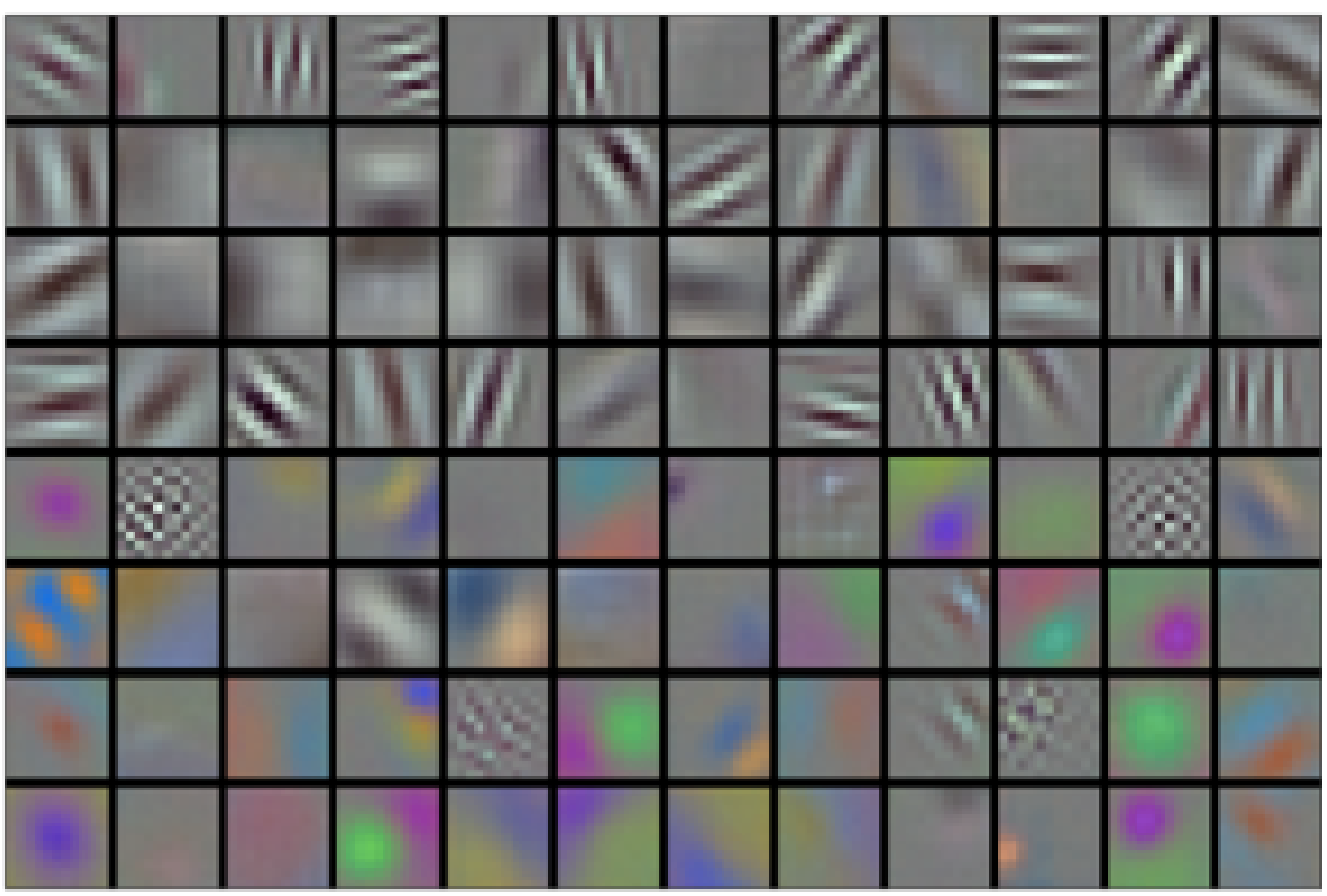

Fig. 8.1.1 Image filters learned by the first layer of AlexNet. Reproduction courtesy of Krizhevsky et al. (2012).

Interestingly, in the lowest layers of the network, the model learned feature extractors that resembled some traditional filters. Fig. 8.1.1 shows lower-level image descriptors. Higher layers in the network might build upon these representations to represent larger structures, like eyes, noses, blades of grass, and so on. Even higher layers might represent whole objects like people, airplanes, dogs, or frisbees. Ultimately, the final hidden state learns a compact representation of the image that summarizes its contents such that data belonging to different categories can be easily separated.

AlexNet (2012) and its precursor LeNet (1995) share many architectural elements. This begs the question: why did it take so long? A key difference was that, over the previous two decades, the amount of data and the computing power available had increased significantly. As such AlexNet was much larger: it was trained on much more data, and on much faster GPUs compared to the CPUs available in 1995.

### Missing Ingredient: Data

Deep models with many layers require large amounts of data in order to enter the regime where they significantly outperform traditional methods based on convex optimizations (e.g., linear and kernel methods). However, given the limited storage capacity of computers, the relative expense of (imaging) sensors, and the comparatively tighter research budgets in the 1990s, most research relied on tiny datasets. Numerous papers relied on the UCI collection of datasets, many of which contained only hundreds or (a few) thousands of images captured in low resolution and often with an artificially clean background.

In 2009, the ImageNet dataset was released (Deng *et al.*, 2009), challenging researchers to learn models from 1 million examples, 1000 each from 1000 distinct categories of objects. The categories themselves were based on the most popular noun nodes in WordNet (Miller, 1995). The ImageNet team used Google Image Search to prefilter large candidate sets for each category and employed the Amazon Mechanical Turk crowdsourcing pipeline to confirm for each image whether it belonged to the associated category. This scale was unprecedented, exceeding others by over an order of magnitude (e.g., CIFAR-100 has 60,000 images). Another aspect was that the images were at relatively high resolution of $224 \times 224$ pixels, unlike the 80 million-sized TinyImages dataset (Torralba *et al.*, 2008), consisting of $32 \times 32$ pixel thumbnails. This allowed for the formation of higher-level features. The associated competition, dubbed the ImageNet Large Scale Visual Recognition Challenge (Russakovsky *et al.*, 2015), pushed computer vision and machine learning research forward, challenging researchers to identify which models performed best at a greater scale than academics had previously considered. The largest vision datasets, such as LAION-5B (Schuhmann *et al.*, 2022) contain billions of images with additional metadata.

### Missing Ingredient: Hardware

Deep learning models are voracious consumers of compute cycles. Training can take hundreds of epochs, and each iteration requires passing data through many layers of computationally expensive linear algebra operations. This is one of the main reasons why in the 1990s and early 2000s, simple algorithms based on the more-efficiently optimized convex objectives were preferred.

*Graphical processing units* (GPUs) proved to be a game changer in making deep learning feasible. These chips had earlier been developed for accelerating graphics processing to benefit computer games. In particular, they were optimized for high throughput $4 \times 4$ matrix–vector products, which are needed for many computer graphics tasks. Fortunately, the math is strikingly similar to that required for calculating convolutional layers. Around that time, NVIDIA and ATI had begun optimizing GPUs for general computing operations (Fernando, 2004), going as far as to market them as *general-purpose GPUs* (GPGPUs).

To provide some intuition, consider the cores of a modern microprocessor (CPU). Each of the cores is fairly powerful running at a high clock frequency and sporting large caches (up to several megabytes of L3). Each core is well-suited to executing a wide range of instructions, with branch predictors, a deep pipeline, specialized execution units, speculative execution, and many other bells and whistles that enable it to run a large variety of programs with sophisticated control flow. This apparent strength, however, is also its Achilles heel: general-purpose cores are very expensive to build. They excel at general-purpose code with lots of control flow. This requires lots of chip area, not just for the actual ALU (arithmetic logical unit) where computation happens, but also for all the aforementioned bells and whistles, plus memory interfaces, caching logic between cores, high-speed interconnects, and so on. CPUs are comparatively bad at any single task when compared with dedicated hardware. Modern laptops have 4–8 cores, and even high-end servers rarely exceed 64 cores per socket, simply because it is not cost-effective.

By comparison, GPUs can consist of thousands of small processing elements (NIVIDA's latest Ampere chips have up to 6912 CUDA cores), often grouped into larger groups (NVIDIA calls them warps). The details differ somewhat between NVIDIA, AMD, ARM and other chip vendors. While each core is relatively weak, running at about 1GHz clock frequency, it is the total number of such cores that makes GPUs orders of magnitude faster than CPUs. For instance, NVIDIA's recent Ampere A100 GPU offers over 300 TFLOPs per chip for specialized 16-bit precision (BFLOAT16) matrix-matrix multiplications, and up to 20 TFLOPs for more general-purpose floating point operations (FP32). At the same time, floating point performance of CPUs rarely exceeds 1 TFLOPs. For instance, Amazon's Graviton 3 reaches 2 TFLOPs peak performance for 16-bit precision operations, a number similar to the GPU performance of Apple's M1 processor.

There are many reasons why GPUs are much faster than CPUs in terms of FLOPs. First, power consumption tends to grow *quadratically* with clock frequency. Hence, for the power budget of a CPU core that runs four times faster (a typical number), you can use 16 GPU cores at $\frac{1}{4}$ the speed, which yields $16 \times \frac{1}{4} = 4$ times the performance. Second, GPU cores are much simpler (in fact, for a long time they were not even *able* to execute general-purpose code), which makes them more energy efficient. For instance, (i) they tend not to support speculative evaluation, (ii) it typically is not possible to program each processing element individually, and (iii) the caches per core tend to be much smaller. Last, many operations in deep learning require high memory bandwidth. Again, GPUs shine here with buses that are at least 10 times as wide as many CPUs.

Back to 2012. A major breakthrough came when Alex Krizhevsky and Ilya Sutskever implemented a deep CNN that could run on GPUs. They realized that the computational bottlenecks in CNNs, convolutions and matrix multiplications, are all operations that could be parallelized in hardware. Using two NVIDIA GTX 580s with 3GB of memory, either of which was capable of 1.5 TFLOPs (still a challenge for most CPUs a decade later), they implemented fast convolutions. The cuda-convnet[127] code was good enough that for several years it was the industry standard and powered the first couple of years of the deep learning boom.

### 8.1.2 AlexNet

AlexNet, which employed an 8-layer CNN, won the ImageNet Large Scale Visual Recognition Challenge 2012 by a large margin (Russakovsky *et al.*, 2013). This network showed, for the first time, that the features obtained by learning can transcend manually-designed features, breaking the previous paradigm in computer vision.

The architectures of AlexNet and LeNet are strikingly similar, as Fig. 8.1.2 illustrates. Note that we provide a slightly streamlined version of AlexNet removing some of the design quirks that were needed in 2012 to make the model fit on two small GPUs.

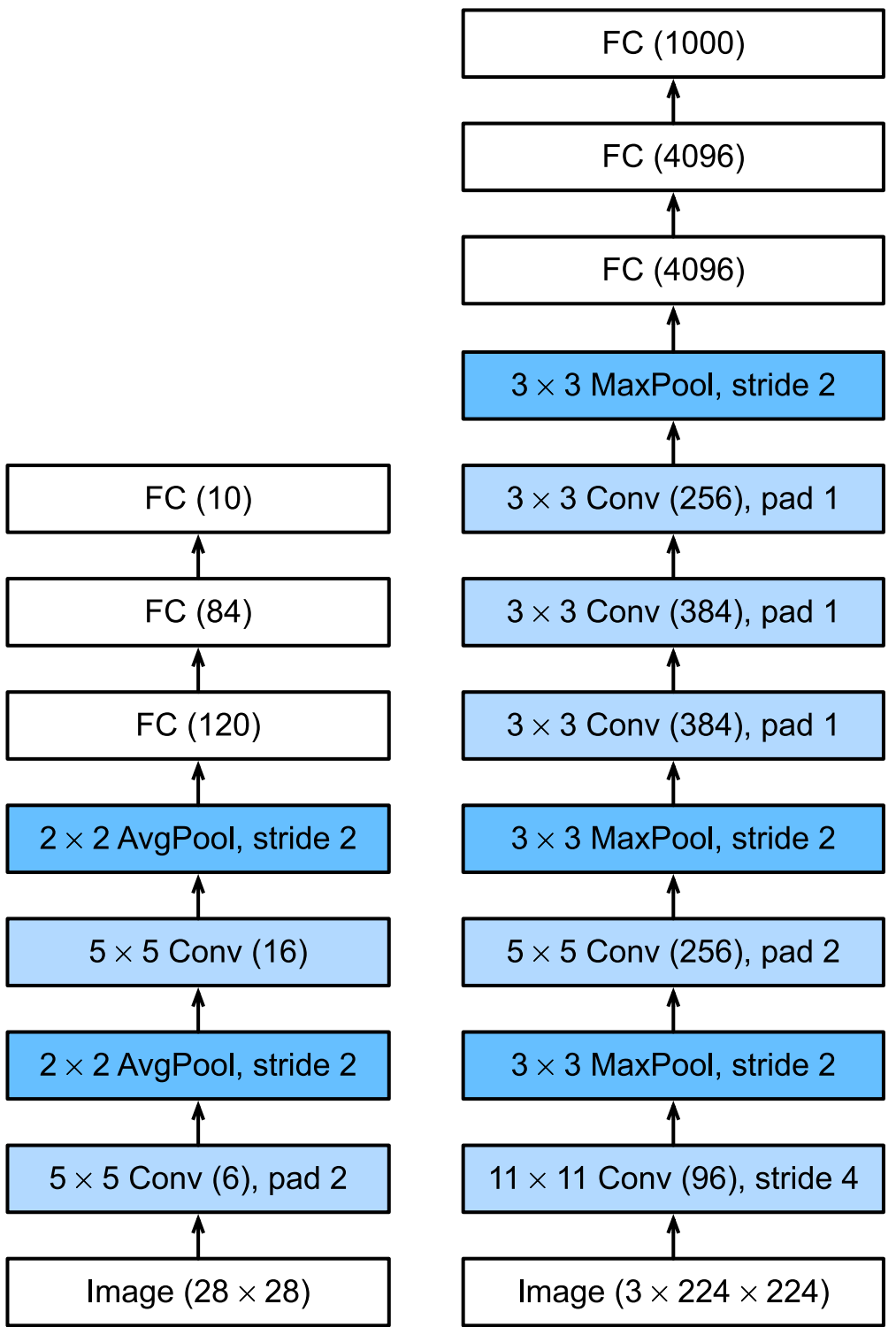

Fig. 8.1.2 From LeNet (left) to AlexNet (right).

There are also significant differences between AlexNet and LeNet. First, AlexNet is much deeper than the comparatively small LeNet-5. AlexNet consists of eight layers: five convolutional layers, two fully connected hidden layers, and one fully connected output layer. Second, AlexNet used the ReLU instead of the sigmoid as its activation function. Let's delve into the details below.

### Architecture

In AlexNet's first layer, the convolution window shape is $11 \times 11$. Since the images in ImageNet are eight times taller and wider than the MNIST images, objects in ImageNet data tend to occupy more pixels with more visual detail. Consequently, a larger convolution window is needed to capture the object. The convolution window shape in the second layer is reduced to $5 \times 5$, followed by $3 \times 3$. In addition, after the first, second, and fifth convolutional layers, the network adds max-pooling layers with a window shape of $3 \times 3$ and a stride of 2. Moreover, AlexNet has ten times more convolution channels than LeNet.

After the final convolutional layer, there are two huge fully connected layers with 4096 outputs. These layers require nearly 1GB model parameters. Because of the limited memory in early GPUs, the original AlexNet used a dual data stream design, so that each of their two GPUs could be responsible for storing and computing only its half of the model. Fortunately, GPU memory is comparatively abundant now, so we rarely need to break up models across GPUs these days (our version of the AlexNet model deviates from the original paper in this aspect).

### Activation Functions

Furthermore, AlexNet changed the sigmoid activation function to a simpler ReLU activation function. On the one hand, the computation of the ReLU activation function is simpler. For example, it does not have the exponentiation operation found in the sigmoid activation function. On the other hand, the ReLU activation function makes model training easier when using different parameter initialization methods. This is because, when the output of the sigmoid activation function is very close to 0 or 1, the gradient of these regions is almost 0, so that backpropagation cannot continue to update some of the model parameters. By contrast, the gradient of the ReLU activation function in the positive interval is always 1 (Section 5.1.2). Therefore, if the model parameters are not properly initialized, the sigmoid function may obtain a gradient of almost 0 in the positive interval, meaning that the model cannot be effectively trained.

### Capacity Control and Preprocessing

AlexNet controls the model complexity of the fully connected layer by dropout (Section 5.6), while LeNet only uses weight decay. To augment the data even further, the training loop of AlexNet added a great deal of image augmentation, such as flipping, clipping, and color changes. This makes the model more robust and the larger sample size effectively reduces overfitting. See Buslaev *et al.* (2020) for an in-depth review of such preprocessing steps.

```
class AlexNet(d2l.Classifier):
    def __init__(self, lr=0.1, num_classes=10):
        super().__init__()
        self.save_hyperparameters()
        self.net = nn.Sequential(
            nn.LazyConv2d(96, kernel_size=11, stride=4, padding=1),
            nn.ReLU(), nn.MaxPool2d(kernel_size=3, stride=2),
            nn.LazyConv2d(256, kernel_size=5, padding=2), nn.ReLU(),
```

```
            nn.MaxPool2d(kernel_size=3, stride=2),
            nn.LazyConv2d(384, kernel_size=3, padding=1), nn.ReLU(),
            nn.LazyConv2d(384, kernel_size=3, padding=1), nn.ReLU(),
            nn.LazyConv2d(256, kernel_size=3, padding=1), nn.ReLU(),
            nn.MaxPool2d(kernel_size=3, stride=2), nn.Flatten(),
            nn.LazyLinear(4096), nn.ReLU(), nn.Dropout(p=0.5),
            nn.LazyLinear(4096), nn.ReLU(),nn.Dropout(p=0.5),
            nn.LazyLinear(num_classes))
        self.net.apply(d2l.init_cnn)
```

We construct a single-channel data example with both height and width of 224 to observe the output shape of each layer. It matches the AlexNet architecture in Fig. 8.1.2.

```
AlexNet().layer_summary((1, 1, 224, 224))
```

```
Conv2d output shape:         torch.Size([1, 96, 54, 54])
ReLU output shape:   torch.Size([1, 96, 54, 54])
MaxPool2d output shape:      torch.Size([1, 96, 26, 26])
Conv2d output shape:         torch.Size([1, 256, 26, 26])
ReLU output shape:   torch.Size([1, 256, 26, 26])
MaxPool2d output shape:      torch.Size([1, 256, 12, 12])
Conv2d output shape:         torch.Size([1, 384, 12, 12])
ReLU output shape:   torch.Size([1, 384, 12, 12])
Conv2d output shape:         torch.Size([1, 384, 12, 12])
ReLU output shape:   torch.Size([1, 384, 12, 12])
Conv2d output shape:         torch.Size([1, 256, 12, 12])
ReLU output shape:   torch.Size([1, 256, 12, 12])
MaxPool2d output shape:      torch.Size([1, 256, 5, 5])
Flatten output shape:        torch.Size([1, 6400])
Linear output shape:         torch.Size([1, 4096])
ReLU output shape:   torch.Size([1, 4096])
Dropout output shape:        torch.Size([1, 4096])
Linear output shape:         torch.Size([1, 4096])
ReLU output shape:   torch.Size([1, 4096])
Dropout output shape:        torch.Size([1, 4096])
Linear output shape:         torch.Size([1, 10])
```

## 8.1.3 Training

Although AlexNet was trained on ImageNet in Krizhevsky *et al.* (2012), we use Fashion-MNIST here since training an ImageNet model to convergence could take hours or days even on a modern GPU. One of the problems with applying AlexNet directly on Fashion-MNIST is that its images have lower resolution ($28\times28$ pixels) than ImageNet images. To make things work, we upsample them to $224 \times 224$. This is generally not a smart practice, as it simply increases the computational complexity without adding information. Nonetheless, we do it here to be faithful to the AlexNet architecture. We perform this resizing with the `resize` argument in the `d2l.FashionMNIST` constructor.

Now, we can start training AlexNet. Compared to LeNet in Section 7.6, the main change

here is the use of a smaller learning rate and much slower training due to the deeper and wider network, the higher image resolution, and the more costly convolutions.

```
model = AlexNet(lr=0.01)
data = d2l.FashionMNIST(batch_size=128, resize=(224, 224))
trainer = d2l.Trainer(max_epochs=10, num_gpus=1)
trainer.fit(model, data)
```

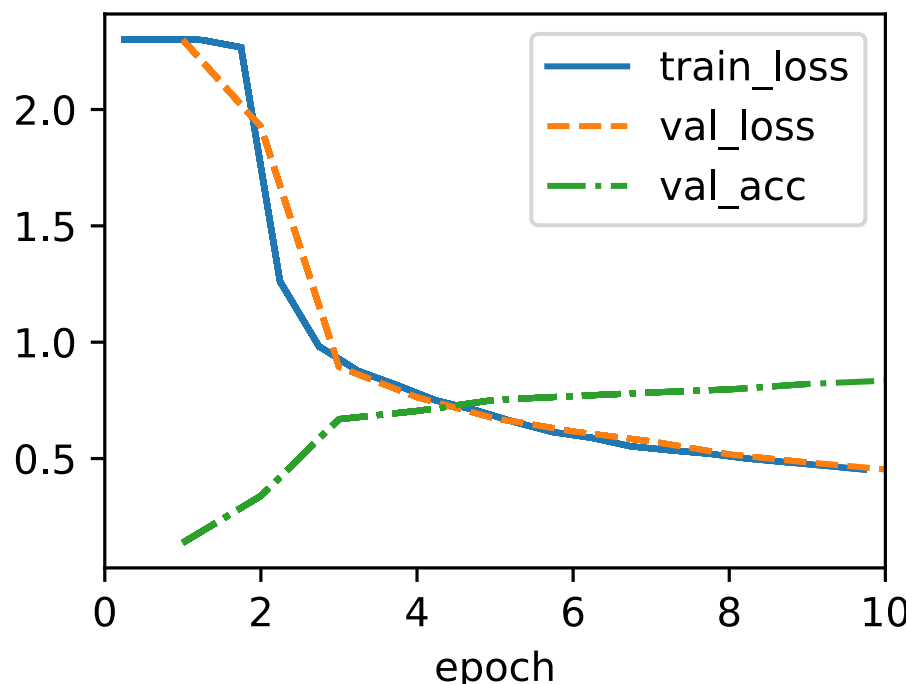

## 8.1.4 Discussion

AlexNet's structure bears a striking resemblance to LeNet, with a number of critical improvements, both for accuracy (dropout) and for ease of training (ReLU). What is equally striking is the amount of progress that has been made in terms of deep learning tooling. What was several months of work in 2012 can now be accomplished in a dozen lines of code using any modern framework.

Reviewing the architecture, we see that AlexNet has an Achilles heel when it comes to efficiency: the last two hidden layers require matrices of size $6400 \times 4096$ and $4096 \times 4096$, respectively. This corresponds to 164 MB of memory and 81 MFLOPs of computation, both of which are a nontrivial outlay, especially on smaller devices, such as mobile phones. This is one of the reasons why AlexNet has been surpassed by much more effective architectures that we will cover in the following sections. Nonetheless, it is a key step from shallow to deep networks that are used nowadays. Note that even though the number of parameters exceeds by far the amount of training data in our experiments (the last two layers have more than 40 million parameters, trained on a datasets of 60 thousand images), there is hardly any overfitting: training and validation loss are virtually identical throughout training. This is due to the improved regularization, such as dropout, inherent in modern deep network designs.

Although it seems that there are only a few more lines in AlexNet's implementation than in LeNet's, it took the academic community many years to embrace this conceptual change and take advantage of its excellent experimental results. This was also due to the lack of efficient computational tools. At the time neither DistBelief (Dean *et al.*, 2012) nor Caffe (Jia *et al.*, 2014) existed, and Theano (Bergstra *et al.*, 2010) still lacked many distinguishing features. It was the availability of TensorFlow (Abadi *et al.*, 2016) that dramatically changed the situation.

### 8.1.5 Exercises

1. Following up on the discussion above, analyze the computational properties of AlexNet.
    1. Compute the memory footprint for convolutions and fully connected layers, respectively. Which one dominates?
    2. Calculate the computational cost for the convolutions and the fully connected layers.
    3. How does the memory (read and write bandwidth, latency, size) affect computation? Is there any difference in its effects for training and inference?
2. You are a chip designer and need to trade off computation and memory bandwidth. For example, a faster chip requires more power and possibly a larger chip area. More memory bandwidth requires more pins and control logic, thus also more area. How do you optimize?
3. Why do engineers no longer report performance benchmarks on AlexNet?
4. Try increasing the number of epochs when training AlexNet. Compared with LeNet, how do the results differ? Why?
5. AlexNet may be too complex for the Fashion-MNIST dataset, in particular due to the low resolution of the initial images.
    1. Try simplifying the model to make the training faster, while ensuring that the accuracy does not drop significantly.
    2. Design a better model that works directly on $28 \times 28$ images.
6. Modify the batch size, and observe the changes in throughput (images/s), accuracy, and GPU memory.
7. Apply dropout and ReLU to LeNet-5. Does it improve? Can you improve things further by preprocessing to take advantage of the invariances inherent in the images?
8. Can you make AlexNet overfit? Which feature do you need to remove or change to break training?

Discussions[128].

## 8.2 Networks Using Blocks (VGG)

While AlexNet offered empirical evidence that deep CNNs can achieve good results, it did not provide a general template to guide subsequent researchers in designing new networks. In the following sections, we will introduce several heuristic concepts commonly used to design deep networks.

Progress in this field mirrors that of VLSI (very large scale integration) in chip design where engineers moved from placing transistors to logical elements to logic blocks (Mead, 1980). Similarly, the design of neural network architectures has grown progressively more abstract, with researchers moving from thinking in terms of individual neurons to whole layers, and now to blocks, repeating patterns of layers. A decade later, this has now progressed to researchers using entire trained models to repurpose them for different, albeit related, tasks. Such large pretrained models are typically called *foundation models* (Bommasani *et al.*, 2021).

Back to network design. The idea of using blocks first emerged from the Visual Geometry Group (VGG) at Oxford University, in their eponymously-named *VGG* network (Simonyan and Zisserman, 2014). It is easy to implement these repeated structures in code with any modern deep learning framework by using loops and subroutines.

```
import torch
from torch import nn
from d2l import torch as d2l
```

### 8.2.1 VGG Blocks

The basic building block of CNNs is a sequence of the following: (i) a convolutional layer with padding to maintain the resolution, (ii) a nonlinearity such as a ReLU, (iii) a pooling layer such as max-pooling to reduce the resolution. One of the problems with this approach is that the spatial resolution decreases quite rapidly. In particular, this imposes a hard limit of $\log_2 d$ convolutional layers on the network before all dimensions ($d$) are used up. For instance, in the case of ImageNet, it would be impossible to have more than 8 convolutional layers in this way.

The key idea of Simonyan and Zisserman (2014) was to use *multiple* convolutions in between downsampling via max-pooling in the form of a block. They were primarily interested in whether deep or wide networks perform better. For instance, the successive application of two $3\times3$ convolutions touches the same pixels as a single $5\times5$ convolution does. At the same time, the latter uses approximately as many parameters ($25\cdot c^2$) as three $3\times3$ convolutions do ($3\cdot9\cdot c^2$). In a rather detailed analysis they showed that deep and narrow networks significantly outperform their shallow counterparts. This set deep learning on a quest for ever deeper networks with over 100 layers for typical applications. Stacking $3\times3$ convolutions has become a gold standard in later deep networks (a design decision only to be revisited recently by Liu *et* *al.* (2022)). Consequently, fast implementations for small convolutions have become a staple on GPUs (Lavin and Gray, 2016).

Back to VGG: a VGG block consists of a *sequence* of convolutions with $3 \times 3$ kernels with padding of 1 (keeping height and width) followed by a $2 \times 2$ max-pooling layer with stride of 2 (halving height and width after each block). In the code below, we define a function called `vgg_block` to implement one VGG block.

The function below takes two arguments, corresponding to the number of convolutional layers `num_convs` and the number of output channels `num_channels`.

```
def vgg_block(num_convs, out_channels):
    layers = []
    for _ in range(num_convs):
        layers.append(nn.LazyConv2d(out_channels, kernel_size=3, padding=1))
        layers.append(nn.ReLU())
    layers.append(nn.MaxPool2d(kernel_size=2,stride=2))
    return nn.Sequential(*layers)
```

### 8.2.2 VGG Network

Like AlexNet and LeNet, the VGG Network can be partitioned into two parts: the first consisting mostly of convolutional and pooling layers and the second consisting of fully connected layers that are identical to those in AlexNet. The key difference is that the convolutional layers are grouped in nonlinear transformations that leave the dimensonality unchanged, followed by a resolution-reduction step, as depicted in Fig. 8.2.1.

The convolutional part of the network connects several VGG blocks from Fig. 8.2.1 (also defined in the `vgg_block` function) in succession. This grouping of convolutions is a pattern that has remained almost unchanged over the past decade, although the specific choice of operations has undergone considerable modifications. The variable `arch` consists of a list of tuples (one per block), where each contains two values: the number of convolutional layers and the number of output channels, which are precisely the arguments required to call the `vgg_block` function. As such, VGG defines a *family* of networks rather than just a specific manifestation. To build a specific network we simply iterate over `arch` to compose the blocks.

```
class VGG(d2l.Classifier):
    def __init__(self, arch, lr=0.1, num_classes=10):
        super().__init__()
        self.save_hyperparameters()
        conv_blks = []
        for (num_convs, out_channels) in arch:
            conv_blks.append(vgg_block(num_convs, out_channels))
        self.net = nn.Sequential(
            *conv_blks, nn.Flatten(),
            nn.LazyLinear(4096), nn.ReLU(), nn.Dropout(0.5),
            nn.LazyLinear(4096), nn.ReLU(), nn.Dropout(0.5),
```

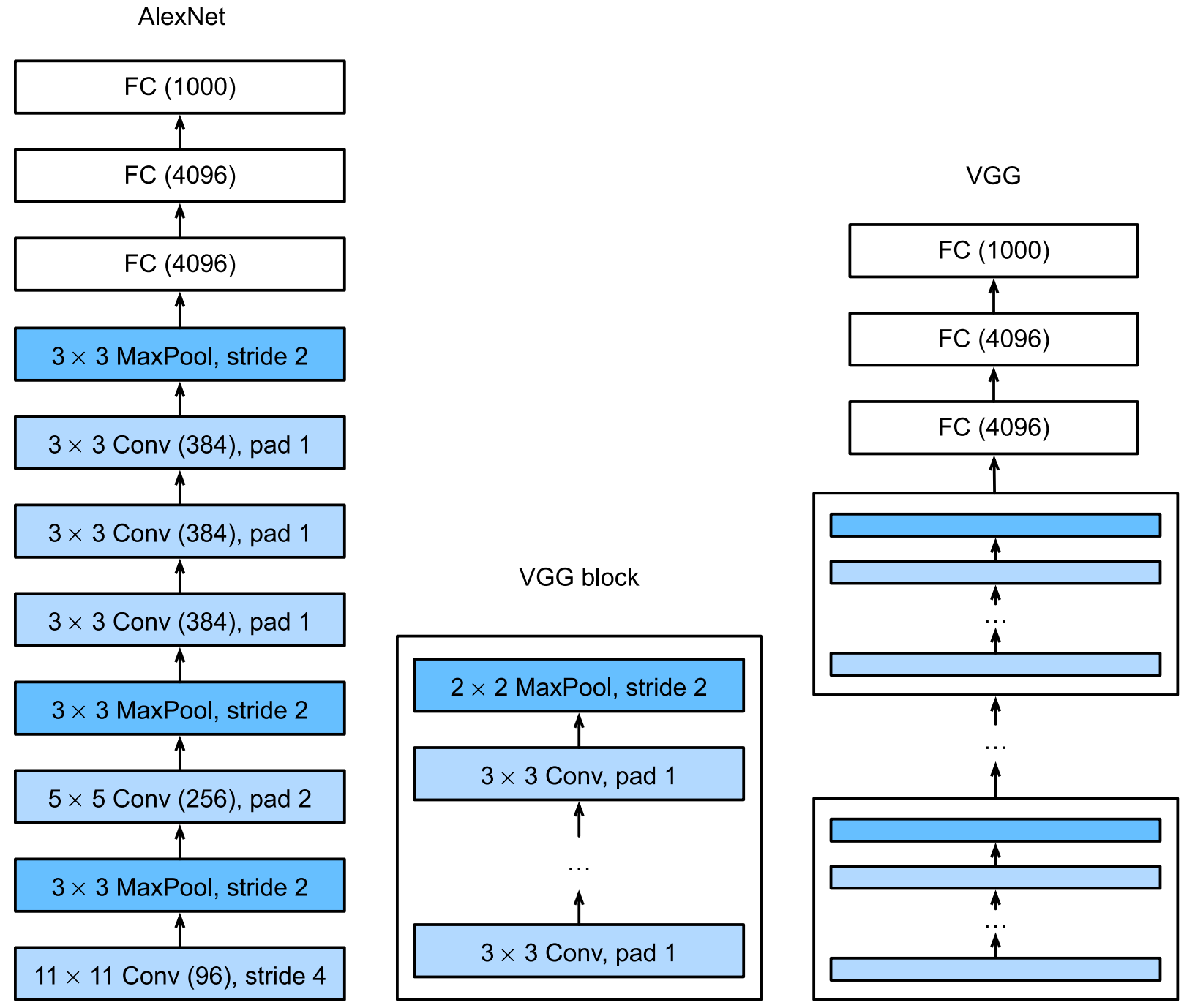

Fig. 8.2.1 From AlexNet to VGG. The key difference is that VGG consists of blocks of layers, whereas AlexNet's layers are all designed individually.

```
            nn.LazyLinear(num_classes))
        self.net.apply(d2l.init_cnn)
```

The original VGG network had five convolutional blocks, among which the first two have one convolutional layer each and the latter three contain two convolutional layers each. The first block has 64 output channels and each subsequent block doubles the number of output channels, until that number reaches 512. Since this network uses eight convolutional layers and three fully connected layers, it is often called VGG-11.

```
VGG(arch=((1, 64), (1, 128), (2, 256), (2, 512), (2, 512))).layer_summary(
    (1, 1, 224, 224))
```

```
Sequential output shape:     torch.Size([1, 64, 112, 112])
Sequential output shape:     torch.Size([1, 128, 56, 56])
Sequential output shape:     torch.Size([1, 256, 28, 28])
Sequential output shape:     torch.Size([1, 512, 14, 14])
Sequential output shape:     torch.Size([1, 512, 7, 7])
Flatten output shape:        torch.Size([1, 25088])
Linear output shape:         torch.Size([1, 4096])
ReLU output shape:   torch.Size([1, 4096])
```

```
Dropout output shape:        torch.Size([1, 4096])
Linear output shape:         torch.Size([1, 4096])
ReLU output shape:   torch.Size([1, 4096])
Dropout output shape:        torch.Size([1, 4096])
Linear output shape:         torch.Size([1, 10])
```

As you can see, we halve height and width at each block, finally reaching a height and width of 7 before flattening the representations for processing by the fully connected part of the network. Simonyan and Zisserman (2014) described several other variants of VGG. In fact, it has become the norm to propose *families* of networks with different speed–accuracy trade-off when introducing a new architecture.

### 8.2.3 Training

Since VGG-11 is computationally more demanding than AlexNet we construct a network with a smaller number of channels. This is more than sufficient for training on Fashion-MNIST. The model training process is similar to that of AlexNet in Section 8.1. Again observe the close match between validation and training loss, suggesting only a small amount of overfitting.

```
model = VGG(arch=((1, 16), (1, 32), (2, 64), (2, 128), (2, 128)), lr=0.01)
trainer = d2l.Trainer(max_epochs=10, num_gpus=1)
data = d2l.FashionMNIST(batch_size=128, resize=(224, 224))
model.apply_init([next(iter(data.get_dataloader(True)))[0]], d2l.init_cnn)
trainer.fit(model, data)
```

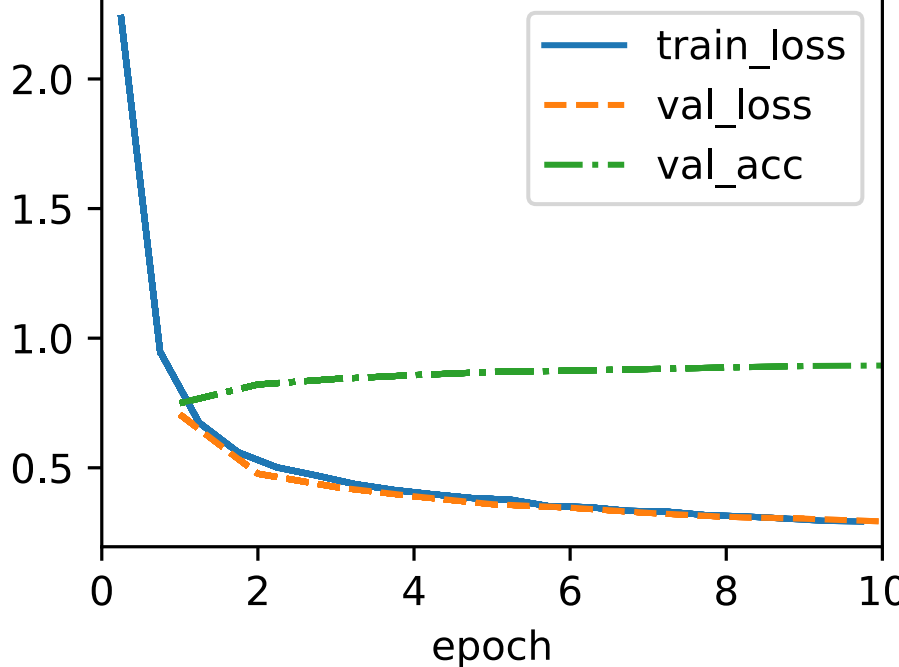

### 8.2.4 Summary

One might argue that VGG is the first truly modern convolutional neural network. While AlexNet introduced many of the components of what make deep learning effective at scale, it is VGG that arguably introduced key properties such as blocks of multiple convolutions and a preference for deep and narrow networks. It is also the first network that is actually an entire family of similarly parametrized models, giving the practitioner ample trade-off between complexity and speed. This is also the place where modern deep learning frameworks shine. It is no longer necessary to generate XML configuration files to specify a network but rather, to assemble said networks through simple Python code.

More recently ParNet (Goyal *et al.*, 2021) demonstrated that it is possible to achieve competitive performance using a much more shallow architecture through a large number of parallel computations. This is an exciting development and there is hope that it will influence architecture designs in the future. For the remainder of the chapter, though, we will follow the path of scientific progress over the past decade.

### 8.2.5 Exercises

1. Compared with AlexNet, VGG is much slower in terms of computation, and it also needs more GPU memory.
    1. Compare the number of parameters needed for AlexNet and VGG.
    2. Compare the number of floating point operations used in the convolutional layers and in the fully connected layers.
    3. How could you reduce the computational cost created by the fully connected layers?
2. When displaying the dimensions associated with the various layers of the network, we only see the information associated with eight blocks (plus some auxiliary transforms), even though the network has 11 layers. Where did the remaining three layers go?
3. Use Table 1 in the VGG paper (Simonyan and Zisserman, 2014) to construct other common models, such as VGG-16 or VGG-19.
4. Upsampling the resolution in Fashion-MNIST eight-fold from $28\times28$ to $224\times224$ dimensions is very wasteful. Try modifying the network architecture and resolution conversion, e.g., to 56 or to 84 dimensions for its input instead. Can you do so without reducing the accuracy of the network? Consult the VGG paper (Simonyan and Zisserman, 2014) for ideas on adding more nonlinearities prior to downsampling.

Discussions[129].

## 8.3 Network in Network (NiN)

LeNet, AlexNet, and VGG all share a common design pattern: extract features exploiting *spatial* structure via a sequence of convolutions and pooling layers and post-process the rep-resentations via fully connected layers. The improvements upon LeNet by AlexNet and VGG mainly lie in how these later networks widen and deepen these two modules.

This design poses two major challenges. First, the fully connected layers at the end of the architecture consume tremendous numbers of parameters. For instance, even a simple model such as VGG-11 requires a monstrous matrix, occupying almost 400MB of RAM in single precision (FP32). This is a significant impediment to computation, in particular on mobile and embedded devices. After all, even high-end mobile phones sport no more than 8GB of RAM. At the time VGG was invented, this was an order of magnitude less (the iPhone 4S had 512MB). As such, it would have been difficult to justify spending the majority of memory on an image classifier.

Second, it is equally impossible to add fully connected layers earlier in the network to increase the degree of nonlinearity: doing so would destroy the spatial structure and require potentially even more memory.

The *network in network* (*NiN*) blocks (Lin *et al.*, 2013) offer an alternative, capable of solving both problems in one simple strategy. They were proposed based on a very simple insight: (i) use $1 \times 1$ convolutions to add local nonlinearities across the channel activations and (ii) use global average pooling to integrate across all locations in the last representation layer. Note that global average pooling would not be effective, were it not for the added nonlinearities. Let's dive into this in detail.

```
import torch
from torch import nn
from d2l import torch as d2l
```

## 8.3.1 NiN Blocks

Recall Section 7.4.3. In it we said that the inputs and outputs of convolutional layers consist of four-dimensional tensors with axes corresponding to the example, channel, height, and width. Also recall that the inputs and outputs of fully connected layers are typically two-dimensional tensors corresponding to the example and feature. The idea behind NiN is to apply a fully connected layer at each pixel location (for each height and width). The resulting $1 \times 1$ convolution can be thought of as a fully connected layer acting independently on each pixel location.

Fig. 8.3.1 illustrates the main structural differences between VGG and NiN, and their blocks. Note both the difference in the NiN blocks (the initial convolution is followed by $1 \times 1$ convolutions, whereas VGG retains $3 \times 3$ convolutions) and at the end where we no longer require a giant fully connected layer.

```
def nin_block(out_channels, kernel_size, strides, padding):
    return nn.Sequential(
        nn.LazyConv2d(out_channels, kernel_size, strides, padding), nn.ReLU(),
        nn.LazyConv2d(out_channels, kernel_size=1), nn.ReLU(),
        nn.LazyConv2d(out_channels, kernel_size=1), nn.ReLU())
```

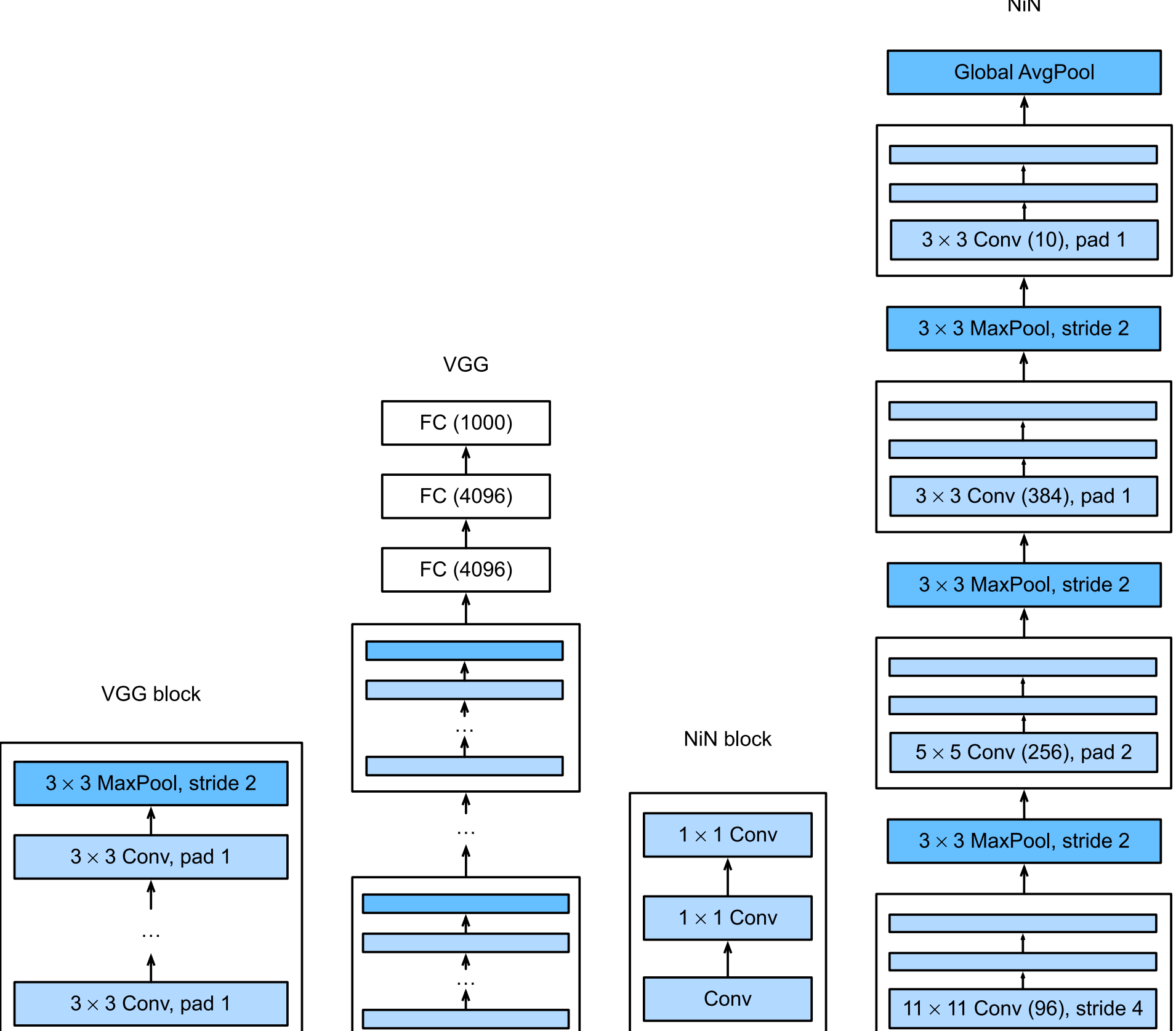

Fig. 8.3.1 Comparing the architectures of VGG and NiN, and of their blocks.

## 8.3.2 NiN Model

NiN uses the same initial convolution sizes as AlexNet (it was proposed shortly thereafter). The kernel sizes are $11\times11$, $5\times5$, and $3\times3$, respectively, and the numbers of output channels match those of AlexNet. Each NiN block is followed by a max-pooling layer with a stride of 2 and a window shape of $3 \times 3$.

The second significant difference between NiN and both AlexNet and VGG is that NiN avoids fully connected layers altogether. Instead, NiN uses a NiN block with a number of output channels equal to the number of label classes, followed by a *global* average pooling layer, yielding a vector of logits. This design significantly reduces the number of required model parameters, albeit at the expense of a potential increase in training time.

```
class NiN(d2l.Classifier):
    def __init__(self, lr=0.1, num_classes=10):
        super().__init__()
        self.save_hyperparameters()
```

```
        self.net = nn.Sequential(
            nin_block(96, kernel_size=11, strides=4, padding=0),
            nn.MaxPool2d(3, stride=2),
            nin_block(256, kernel_size=5, strides=1, padding=2),
            nn.MaxPool2d(3, stride=2),
            nin_block(384, kernel_size=3, strides=1, padding=1),
            nn.MaxPool2d(3, stride=2),
            nn.Dropout(0.5),
            nin_block(num_classes, kernel_size=3, strides=1, padding=1),
            nn.AdaptiveAvgPool2d((1, 1)),
            nn.Flatten())
        self.net.apply(d2l.init_cnn)
```

We create a data example to see the output shape of each block.

```
NiN().layer_summary((1, 1, 224, 224))
```

```
Sequential output shape:     torch.Size([1, 96, 54, 54])
MaxPool2d output shape:      torch.Size([1, 96, 26, 26])
Sequential output shape:     torch.Size([1, 256, 26, 26])
MaxPool2d output shape:      torch.Size([1, 256, 12, 12])
Sequential output shape:     torch.Size([1, 384, 12, 12])
MaxPool2d output shape:      torch.Size([1, 384, 5, 5])
Dropout output shape:        torch.Size([1, 384, 5, 5])
Sequential output shape:     torch.Size([1, 10, 5, 5])
AdaptiveAvgPool2d output shape:      torch.Size([1, 10, 1, 1])
Flatten output shape:        torch.Size([1, 10])
```

## 8.3.3 Training

As before we use Fashion-MNIST to train the model using the same optimizer that we used for AlexNet and VGG.

```
model = NiN(lr=0.05)
trainer = d2l.Trainer(max_epochs=10, num_gpus=1)
data = d2l.FashionMNIST(batch_size=128, resize=(224, 224))
model.apply_init([next(iter(data.get_dataloader(True)))[0]], d2l.init_cnn)
trainer.fit(model, data)
```

## 8.3.4 Summary

NiN has dramatically fewer parameters than AlexNet and VGG. This stems primarily from the fact that it needs no giant fully connected layers. Instead, it uses global average pooling to aggregate across all image locations after the last stage of the network body. This obviates the need for expensive (learned) reduction operations and replaces them by a simple average. What surprised researchers at the time was the fact that this averaging operation did not harm accuracy. Note that averaging across a low-resolution representation (with many channels) also adds to the amount of translation invariance that the network can handle.

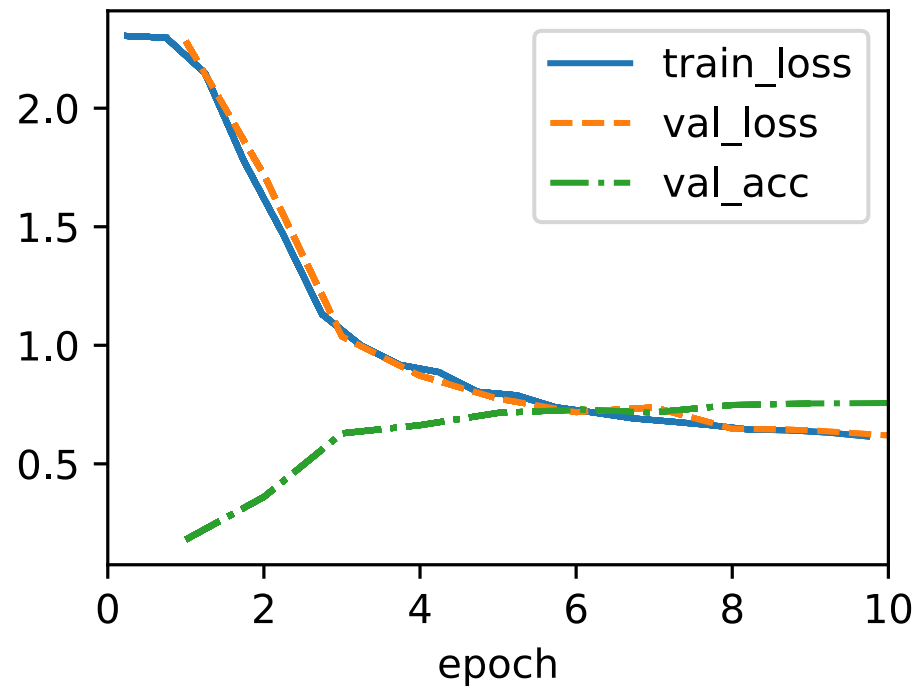

Choosing fewer convolutions with wide kernels and replacing them by $1 \times 1$ convolutions aids the quest for fewer parameters further. It can cater for a significant amount of nonlinearity across channels within any given location. Both $1 \times 1$ convolutions and global average pooling significantly influenced subsequent CNN designs.

### 8.3.5 Exercises

1. Why are there two $1 \times 1$ convolutional layers per NiN block? Increase their number to three. Reduce their number to one. What changes?
2. What changes if you replace the $1 \times 1$ convolutions by $3 \times 3$ convolutions?
3. What happens if you replace the global average pooling by a fully connected layer (speed, accuracy, number of parameters)?
4. Calculate the resource usage for NiN.
    1. What is the number of parameters?
    2. What is the amount of computation?
    3. What is the amount of memory needed during training?
    4. What is the amount of memory needed during prediction?
5. What are possible problems with reducing the $384 \times 5 \times 5$ representation to a $10 \times 5 \times 5$ representation in one step?
6. Use the structural design decisions in VGG that led to VGG-11, VGG-16, and VGG-19 to design a family of NiN-like networks.

Discussions[130].

# 8.4 Multi-Branch Networks (GoogLeNet)

In 2014, *GoogLeNet* won the ImageNet Challenge (Szegedy *et al.*, 2015), using a structure that combined the strengths of NiN (Lin *et al.*, 2013), repeated blocks (Simonyan and Zisserman, 2014), and a cocktail of convolution kernels. It was arguably also the first network that exhibited a clear distinction among the stem (data ingest), body (data processing), and head (prediction) in a CNN. This design pattern has persisted ever since in the design of deep networks: the *stem* is given by the first two or three convolutions that operate on the image. They extract low-level features from the underlying images. This is followed by a *body* of convolutional blocks. Finally, the *head* maps the features obtained so far to the required classification, segmentation, detection, or tracking problem at hand.

The key contribution in GoogLeNet was the design of the network body. It solved the problem of selecting convolution kernels in an ingenious way. While other works tried to identify which convolution, ranging from $1 \times 1$ to $11 \times 11$ would be best, it simply *concatenated* multi-branch convolutions. In what follows we introduce a slightly simplified version of GoogLeNet: the original design included a number of tricks for stabilizing training through intermediate loss functions, applied to multiple layers of the network. They are no longer necessary due to the availability of improved training algorithms.

```
import torch
from torch import nn
from torch.nn import functional as F
from d2l import torch as d2l
```

## 8.4.1 Inception Blocks

The basic convolutional block in GoogLeNet is called an *Inception block*, stemming from the meme "we need to go deeper" from the movie *Inception*.

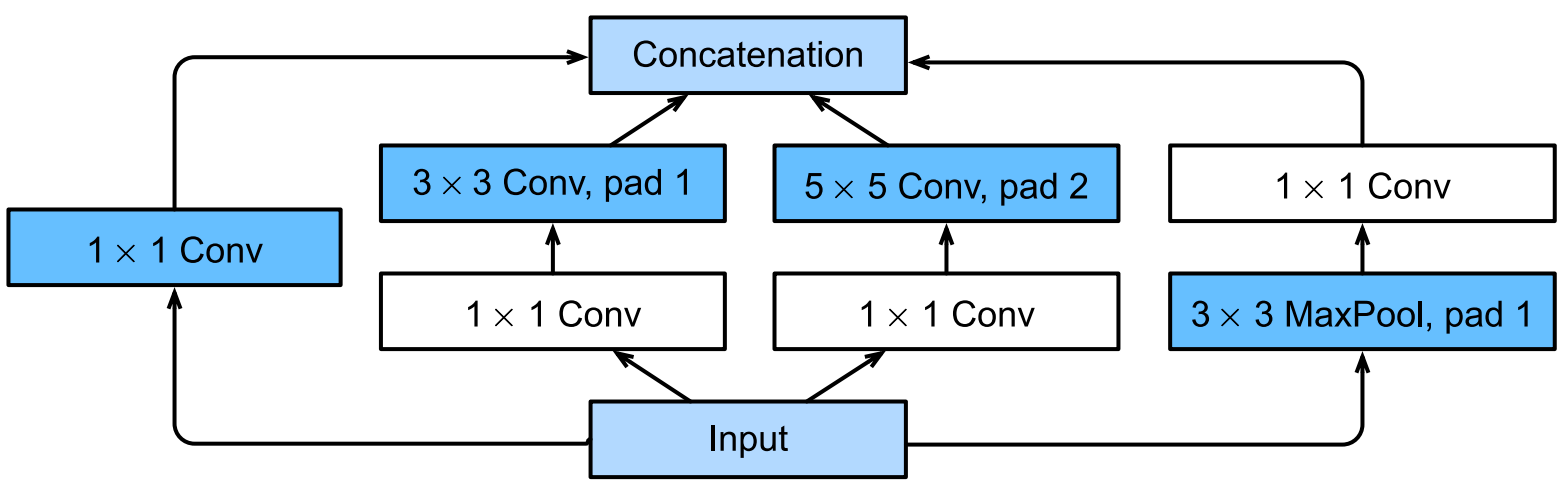

Fig. 8.4.1 Structure of the Inception block.

As depicted in Fig. 8.4.1, the inception block consists of four parallel branches. The first three branches use convolutional layers with window sizes of $1 \times 1$, $3 \times 3$, and $5 \times 5$ to extract information from different spatial sizes. The middle two branches also add a $1 \times 1$ convolution of the input to reduce the number of channels, reducing the model's complexity. The fourth branch uses a $3 \times 3$ max-pooling layer, followed by a $1 \times 1$ convolutional layer to change the number of channels. The four branches all use appropriate padding to give the input and output the same height and width. Finally, the outputs along each branch are concatenated along the channel dimension and comprise the block's output. The commonly-tuned hyperparameters of the Inception block are the number of output channels per layer, i.e., how to allocate capacity among convolutions of different size.

```
class Inception(nn.Module):
    # c1--c4 are the number of output channels for each branch
    def __init__(self, c1, c2, c3, c4, **kwargs):
        super(Inception, self).__init__(**kwargs)
        # Branch 1
        self.b1_1 = nn.LazyConv2d(c1, kernel_size=1)
        # Branch 2
        self.b2_1 = nn.LazyConv2d(c2[0], kernel_size=1)
        self.b2_2 = nn.LazyConv2d(c2[1], kernel_size=3, padding=1)
        # Branch 3
        self.b3_1 = nn.LazyConv2d(c3[0], kernel_size=1)
        self.b3_2 = nn.LazyConv2d(c3[1], kernel_size=5, padding=2)
        # Branch 4
        self.b4_1 = nn.MaxPool2d(kernel_size=3, stride=1, padding=1)
        self.b4_2 = nn.LazyConv2d(c4, kernel_size=1)

    def forward(self, x):
        b1 = F.relu(self.b1_1(x))
        b2 = F.relu(self.b2_2(F.relu(self.b2_1(x))))
        b3 = F.relu(self.b3_2(F.relu(self.b3_1(x))))
        b4 = F.relu(self.b4_2(self.b4_1(x)))
        return torch.cat((b1, b2, b3, b4), dim=1)
```

To gain some intuition for why this network works so well, consider the combination of the filters. They explore the image in a variety of filter sizes. This means that details at different extents can be recognized efficiently by filters of different sizes. At the same time, we can allocate different amounts of parameters for different filters.

## 8.4.2 GoogLeNet Model

As shown in Fig. 8.4.2, GoogLeNet uses a stack of a total of 9 inception blocks, arranged into three groups with max-pooling in between, and global average pooling in its head to generate its estimates. Max-pooling between inception blocks reduces the dimensionality. At its stem, the first module is similar to AlexNet and LeNet.

We can now implement GoogLeNet piece by piece. Let's begin with the stem. The first module uses a 64-channel $7 \times 7$ convolutional layer.

```
class GoogleNet(d2l.Classifier):
    def b1(self):
        return nn.Sequential(
```

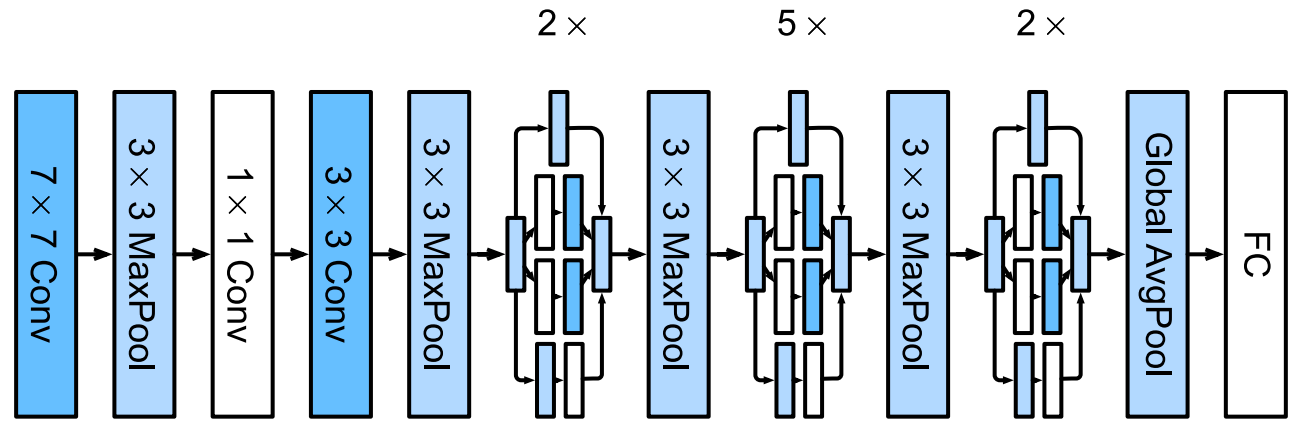

Fig. 8.4.2 The GoogLeNet architecture.

```
            nn.LazyConv2d(64, kernel_size=7, stride=2, padding=3),
            nn.ReLU(), nn.MaxPool2d(kernel_size=3, stride=2, padding=1))
```

The second module uses two convolutional layers: first, a 64-channel $1\times1$ convolutional layer, followed by a $3\times3$ convolutional layer that triples the number of channels. This corresponds to the second branch in the Inception block and concludes the design of the body. At this point we have 192 channels.

```
@d2l.add_to_class(GoogleNet)
def b2(self):
    return nn.Sequential(
        nn.LazyConv2d(64, kernel_size=1), nn.ReLU(),
        nn.LazyConv2d(192, kernel_size=3, padding=1), nn.ReLU(),
        nn.MaxPool2d(kernel_size=3, stride=2, padding=1))
```

The third module connects two complete Inception blocks in series. The number of output channels of the first Inception block is $64+128+32+32=256$. This amounts to a ratio of the number of output channels among the four branches of $2:4:1:1$. To achieve this, we first reduce the input dimensions by $\frac{1}{2}$ and by $\frac{1}{12}$ in the second and third branch respectively to arrive at $96=192/2$ and $16=192/12$ channels respectively.

The number of output channels of the second Inception block is increased to $128+192+96+64=480$, yielding a ratio of $128:192:96:64=4:6:3:2$. As before, we need to reduce the number of intermediate dimensions in the second and third channel. A scale of $\frac{1}{2}$ and $\frac{1}{8}$ respectively suffices, yielding 128 and 32 channels respectively. This is captured by the arguments of the following `Inception` block constructors.

```
@d2l.add_to_class(GoogleNet)
def b3(self):
    return nn.Sequential(Inception(64, (96, 128), (16, 32), 32),
                         Inception(128, (128, 192), (32, 96), 64),
                         nn.MaxPool2d(kernel_size=3, stride=2, padding=1))
```

The fourth module is more complicated. It connects five Inception blocks in series, and they have $192+208+48+64=512$, $160+224+64+64=512$, $128+256+64+64=512$, $112+288+64+64=528$, and $256+320+128+128=832$ output channels, respectively. The number of channels assigned to these branches is similar to that in the third module: the second branch with the $3 \times 3$ convolutional layer outputs the largest number of channels, followed by the first branch with only the $1 \times 1$ convolutional layer, the third branch with the $5 \times 5$ convolutional layer, and the fourth branch with the $3 \times 3$ max-pooling layer. The second and third branches will first reduce the number of channels according to the ratio. These ratios are slightly different in different Inception blocks.

```
@d2l.add_to_class(GoogleNet)
def b4(self):
    return nn.Sequential(Inception(192, (96, 208), (16, 48), 64),
                         Inception(160, (112, 224), (24, 64), 64),
                         Inception(128, (128, 256), (24, 64), 64),
                         Inception(112, (144, 288), (32, 64), 64),
                         Inception(256, (160, 320), (32, 128), 128),
                         nn.MaxPool2d(kernel_size=3, stride=2, padding=1))
```

The fifth module has two Inception blocks with $256 + 320 + 128 + 128 = 832$ and $384 + 384 + 128 + 128 = 1024$ output channels. The number of channels assigned to each branch is the same as that in the third and fourth modules, but differs in specific values. It should be noted that the fifth block is followed by the output layer. This block uses the global average pooling layer to change the height and width of each channel to 1, just as in NiN. Finally, we turn the output into a two-dimensional array followed by a fully connected layer whose number of outputs is the number of label classes.

```
@d2l.add_to_class(GoogleNet)
def b5(self):
    return nn.Sequential(Inception(256, (160, 320), (32, 128), 128),
                         Inception(384, (192, 384), (48, 128), 128),
                         nn.AdaptiveAvgPool2d((1,1)), nn.Flatten())
```

Now that we defined all blocks `b1` through `b5`, it is just a matter of assembling them all into a full network.

```
@d2l.add_to_class(GoogleNet)
def __init__(self, lr=0.1, num_classes=10):
    super(GoogleNet, self).__init__()
    self.save_hyperparameters()
    self.net = nn.Sequential(self.b1(), self.b2(), self.b3(), self.b4(),
                             self.b5(), nn.LazyLinear(num_classes))
    self.net.apply(d2l.init_cnn)
```

The GoogLeNet model is computationally complex. Note the large number of relatively arbitrary hyperparameters in terms of the number of channels chosen, the number of blocks prior to dimensionality reduction, the relative partitioning of capacity across channels, etc. Much of it is due to the fact that at the time when GoogLeNet was introduced, automatic tools for network definition or design exploration were not yet available. For instance, by now we take it for granted that a competent deep learning framework is capable of inferring dimensionalities of input tensors automatically. At the time, many such configurations had to be specified explicitly by the experimenter, thus often slowing down active experimentation. Moreover, the tools needed for automatic exploration were still in flux and initial experiments largely amounted to costly brute-force exploration, genetic algorithms, and similar strategies.

For now the only modification we will carry out is to reduce the input height and width from 224 to 96 to have a reasonable training time on Fashion-MNIST. This simplifies the computation. Let's have a look at the changes in the shape of the output between the various modules.

```
model = GoogleNet().layer_summary((1, 1, 96, 96))
```

```
Sequential output shape:     torch.Size([1, 64, 24, 24])
Sequential output shape:     torch.Size([1, 192, 12, 12])
Sequential output shape:     torch.Size([1, 480, 6, 6])
Sequential output shape:     torch.Size([1, 832, 3, 3])
Sequential output shape:     torch.Size([1, 1024])
Linear output shape:         torch.Size([1, 10])
```

## 8.4.3 Training

As before, we train our model using the Fashion-MNIST dataset. We transform it to $96 \times 96$ pixel resolution before invoking the training procedure.

```
model = GoogleNet(lr=0.01)
trainer = d2l.Trainer(max_epochs=10, num_gpus=1)
data = d2l.FashionMNIST(batch_size=128, resize=(96, 96))
model.apply_init([next(iter(data.get_dataloader(True)))[0]], d2l.init_cnn)
trainer.fit(model, data)
```

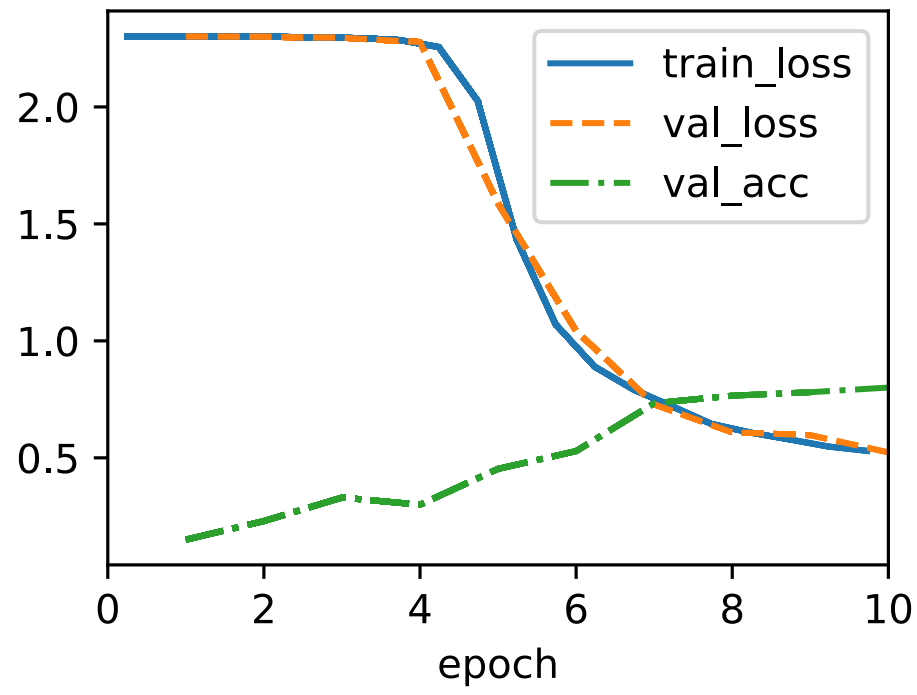

## 8.4.4 Discussion

A key feature of GoogLeNet is that it is actually *cheaper* to compute than its predecessors while simultaneously providing improved accuracy. This marks the beginning of a much more deliberate network design that trades off the cost of evaluating a network with a reduction in errors. It also marks the beginning of experimentation at a block level with network design hyperparameters, even though it was entirely manual at the time. We will revisit this topic in Section 8.8 when discussing strategies for network structure exploration.

Over the following sections we will encounter a number of design choices (e.g., batch normalization, residual connections, and channel grouping) that allow us to improve networks significantly. For now, you can be proud to have implemented what is arguably the first truly modern CNN.

### 8.4.5 Exercises

1. GoogLeNet was so successful that it went through a number of iterations, progressively improving speed and accuracy. Try to implement and run some of them. They include the following:
    1. Add a batch normalization layer (Ioffe and Szegedy, 2015), as described later in Section 8.5.
    2. Make adjustments to the Inception block (width, choice and order of convolutions), as described in Szegedy *et al.* (2016).
    3. Use label smoothing for model regularization, as described in Szegedy *et al.* (2016).
    4. Make further adjustments to the Inception block by adding residual connection (Szegedy *et al.*, 2017), as described later in Section 8.6.
2. What is the minimum image size needed for GoogLeNet to work?
3. Can you design a variant of GoogLeNet that works on Fashion-MNIST's native resolution of $28 \times 28$ pixels? How would you need to change the stem, the body, and the head of the network, if anything at all?
4. Compare the model parameter sizes of AlexNet, VGG, NiN, and GoogLeNet. How do the latter two network architectures significantly reduce the model parameter size?
5. Compare the amount of computation needed in GoogLeNet and AlexNet. How does this affect the design of an accelerator chip, e.g., in terms of memory size, memory bandwidth, cache size, the amount of computation, and the benefit of specialized operations?

Discussions[131].

## 8.5 Batch Normalization

Training deep neural networks is difficult. Getting them to converge in a reasonable amount of time can be tricky. In this section, we describe *batch normalization*, a popular and effective technique that consistently accelerates the convergence of deep networks (Ioffe and Szegedy, 2015). Together with residual blocks—covered later in Section 8.6—batch normalization has made it possible for practitioners to routinely train networks with over 100 layers. A secondary (serendipitous) benefit of batch normalization lies in its inherent regularization.

```
import torch
from torch import nn
from d2l import torch as d2l
```

## 8.5.1 Training Deep Networks

When working with data, we often preprocess before training. Choices regarding data preprocessing often make an enormous difference in the final results. Recall our application of MLPs to predicting house prices (Section 5.7). Our first step when working with real data was to standardize our input features to have zero mean $\boldsymbol{\mu} = 0$ and unit variance $\boldsymbol{\Sigma} = \mathbf{1}$ across multiple observations (Friedman, 1987), frequently rescaling the latter so that the diagonal is unity, i.e., $\Sigma_{ii} = 1$. Yet another strategy is to rescale vectors to unit length, possibly zero mean *per observation*. This can work well, e.g., for spatial sensor data. These preprocessing techniques and many others, are beneficial for keeping the estimation problem well controlled. For a review of feature selection and extraction see the article of Guyon *et al.* (2008), for example. Standardizing vectors also has the nice side-effect of constraining the function complexity of functions that act upon it. For instance, the celebrated radius-margin bound (Vapnik, 1995) in support vector machines and the Perceptron Convergence Theorem (Novikoff, 1962) rely on inputs of bounded norm.

Intuitively, this standardization plays nicely with our optimizers since it puts the parameters *a priori* on a similar scale. As such, it is only natural to ask whether a corresponding normalization step *inside* a deep network might not be beneficial. While this is not quite the reasoning that led to the invention of batch normalization (Ioffe and Szegedy, 2015), it is a useful way of understanding it and its cousin, layer normalization (Ba *et al.*, 2016), within a unified framework.

Second, for a typical MLP or CNN, as we train, the variables in intermediate layers (e.g., affine transformation outputs in MLP) may take values with widely varying magnitudes: whether along the layers from input to output, across units in the same layer, and over time due to our updates to the model parameters. The inventors of batch normalization postulated informally that this drift in the distribution of such variables could hamper the convergence of the network. Intuitively, we might conjecture that if one layer has variable activations that are 100 times that of another layer, this might necessitate compensatory adjustments in the learning rates. Adaptive solvers such as AdaGrad (Duchi *et al.*, 2011), Adam (Kingma and Ba, 2014), Yogi (Zaheer *et al.*, 2018), or Distributed Shampoo (Anil *et al.*, 2020) aim to address this from the viewpoint of optimization, e.g., by adding aspects of second-order methods. The alternative is to prevent the problem from occurring, simply by adaptive normalization.

Third, deeper networks are complex and tend to be more liable to overfitting. This means that regularization becomes more critical. A common technique for regularization is noise injection. This has been known for a long time, e.g., with regard to noise injection for the inputs (Bishop, 1995). It also forms the basis of dropout in Section 5.6. As it turns out, quite serendipitously, batch normalization conveys all three benefits: preprocessing, numerical stability, and regularization.

Batch normalization is applied to individual layers, or optionally, to all of them: In each training iteration, we first normalize the inputs (of batch normalization) by subtracting their mean and dividing by their standard deviation, where both are estimated based on the statistics of the current minibatch. Next, we apply a scale coefficient and an offset to recover the lost degrees of freedom. It is precisely due to this *normalization* based on *batch* statistics that *batch normalization* derives its name.

Note that if we tried to apply batch normalization with minibatches of size 1, we would not be able to learn anything. That is because after subtracting the means, each hidden unit would take value 0. As you might guess, since we are devoting a whole section to batch normalization, with large enough minibatches the approach proves effective and stable. One takeaway here is that when applying batch normalization, the choice of batch size is even more significant than without batch normalization, or at least, suitable calibration is needed as we might adjust batch size.

Denote by $\mathcal{B}$ a minibatch and let $\mathbf{x} \in \mathcal{B}$ be an input to batch normalization (BN). In this case the batch normalization is defined as follows:

$$\mathrm{BN}(\mathbf{x}) = \boldsymbol{\gamma} \odot \frac{\mathbf{x} - \hat{\boldsymbol{\mu}}_\mathcal{B}}{\hat{\boldsymbol{\sigma}}_\mathcal{B}} + \boldsymbol{\beta}. \tag{8.5.1}$$

In (8.5.1), $\hat{\boldsymbol{\mu}}_\mathcal{B}$ is the sample mean and $\hat{\boldsymbol{\sigma}}_\mathcal{B}$ is the sample standard deviation of the minibatch $\mathcal{B}$. After applying standardization, the resulting minibatch has zero mean and unit variance. The choice of unit variance (rather than some other magic number) is arbitrary. We recover this degree of freedom by including an elementwise *scale parameter* $\boldsymbol{\gamma}$ and *shift parameter* $\boldsymbol{\beta}$ that have the same shape as $\mathbf{x}$. Both are parameters that need to be learned as part of model training.

The variable magnitudes for intermediate layers cannot diverge during training since batch normalization actively centers and rescales them back to a given mean and size (via $\hat{\boldsymbol{\mu}}_\mathcal{B}$ and $\hat{\boldsymbol{\sigma}}_\mathcal{B}$). Practical experience confirms that, as alluded to when discussing feature rescaling, batch normalization seems to allow for more aggressive learning rates. We calculate $\hat{\boldsymbol{\mu}}_\mathcal{B}$ and $\hat{\boldsymbol{\sigma}}_\mathcal{B}$ in (8.5.1) as follows:

$$\hat{\boldsymbol{\mu}}_\mathcal{B} = \frac{1}{|\mathcal{B}|} \sum_{\mathbf{x} \in \mathcal{B}} \mathbf{x} \text{ and } \hat{\boldsymbol{\sigma}}_\mathcal{B}^2 = \frac{1}{|\mathcal{B}|} \sum_{\mathbf{x} \in \mathcal{B}} (\mathbf{x} - \hat{\boldsymbol{\mu}}_{\mathcal{B}})^2 + \epsilon. \tag{8.5.2}$$

Note that we add a small constant $\epsilon > 0$ to the variance estimate to ensure that we never attempt division by zero, even in cases where the empirical variance estimate might be very small or vanish. The estimates $\hat{\boldsymbol{\mu}}_\mathcal{B}$ and $\hat{\boldsymbol{\sigma}}_\mathcal{B}$ counteract the scaling issue by using noisy estimates of mean and variance. You might think that this noisiness should be a problem. On the contrary, it is actually beneficial.

This turns out to be a recurring theme in deep learning. For reasons that are not yet well-characterized theoretically, various sources of noise in optimization often lead to faster training and less overfitting: this variation appears to act as a form of regularization. Teye *et al.* (2018) and Luo *et al.* (2018) related the properties of batch normalization to Bayesian priors and penalties, respectively. In particular, this sheds some light on the puzzle of why batch normalization works best for moderate minibatch sizes in the 50–100 range. This particular size of minibatch seems to inject just the "right amount" of noise per layer, both in terms of scale via $\hat{\boldsymbol{\sigma}}$, and in terms of offset via $\hat{\boldsymbol{\mu}}$: a larger minibatch regularizes less due to the more stable estimates, whereas tiny minibatches destroy useful signal due to high variance. Exploring this direction further, considering alternative types of preprocessing and filtering may yet lead to other effective types of regularization.

Fixing a trained model, you might think that we would prefer using the entire dataset to estimate the mean and variance. Once training is complete, why would we want the same image to be classified differently, depending on the batch in which it happens to reside? During training, such exact calculation is infeasible because the intermediate variables for all data examples change every time we update our model. However, once the model is trained, we can calculate the means and variances of each layer's variables based on the entire dataset. Indeed this is standard practice for models employing batch normalization; thus batch normalization layers function differently in *training mode* (normalizing by minibatch statistics) than in *prediction mode* (normalizing by dataset statistics). In this form they closely resemble the behavior of dropout regularization of Section 5.6, where noise is only injected during training.

## 8.5.2 Batch Normalization Layers

Batch normalization implementations for fully connected layers and convolutional layers are slightly different. One key difference between batch normalization and other layers is that because the former operates on a full minibatch at a time, we cannot just ignore the batch dimension as we did before when introducing other layers.

### Fully Connected Layers

When applying batch normalization to fully connected layers, Ioffe and Szegedy (2015), in their original paper inserted batch normalization after the affine transformation and *before* the nonlinear activation function. Later applications experimented with inserting batch normalization right *after* activation functions. Denoting the input to the fully connected layer by $\mathbf{x}$, the affine transformation by $\mathbf{W}\mathbf{x} + \mathbf{b}$ (with the weight parameter $\mathbf{W}$ and the bias parameter $\mathbf{b}$), and the activation function by $\phi$, we can express the computation of a batch-normalization-enabled, fully connected layer output $\mathbf{h}$ as follows:

$$\mathbf{h} = \phi(\mathrm{BN}(\mathbf{W}\mathbf{x} + \mathbf{b})). \tag{8.5.3}$$

Recall that mean and variance are computed on the *same* minibatch on which the transformation is applied.

### Convolutional Layers

Similarly, with convolutional layers, we can apply batch normalization after the convolution but before the nonlinear activation function. The key difference from batch normalization in fully connected layers is that we apply the operation on a per-channel basis *across all locations*. This is compatible with our assumption of translation invariance that led to convolutions: we assumed that the specific location of a pattern within an image was not critical for the purpose of understanding.

Assume that our minibatches contain $m$ examples and that for each channel, the output of the convolution has height $p$ and width $q$. For convolutional layers, we carry out each batch normalization over the $m \cdot p \cdot q$ elements per output channel simultaneously. Thus, we collect the values over all spatial locations when computing the mean and variance and consequently apply the same mean and variance within a given channel to normalize the value at each spatial location. Each channel has its own scale and shift parameters, both of which are scalars.

### Layer Normalization

Note that in the context of convolutions the batch normalization is well defined even for minibatches of size 1: after all, we have all the locations across an image to average. Consequently, mean and variance are well defined, even if it is just within a single observation. This consideration led Ba *et al.* (2016) to introduce the notion of *layer normalization*. It works just like a batch norm, only that it is applied to one observation at a time. Consequently both the offset and the scaling factor are scalars. For an $n$-dimensional vector $\mathbf{x}$, layer norms are given by

$$\mathbf{x} \rightarrow \mathrm{LN}(\mathbf{x}) = \frac{\mathbf{x} - \hat{\mu}}{\hat{\sigma}}, \tag{8.5.4}$$

where scaling and offset are applied coefficient-wise and given by

$$\hat{\mu} \stackrel{\mathrm{def}}{=} \frac{1}{n} \sum_{i=1}^{n} x_i \text{ and } \hat{\sigma}^2 \stackrel{\mathrm{def}}{=} \frac{1}{n} \sum_{i=1}^{n} (x_i - \hat{\mu})^2 + \epsilon. \tag{8.5.5}$$

As before we add a small offset $\epsilon > 0$ to prevent division by zero. One of the major benefits of using layer normalization is that it prevents divergence. After all, ignoring $\epsilon$, the output of the layer normalization is scale independent. That is, we have $\mathrm{LN}(\mathbf{x}) \approx \mathrm{LN}(\alpha\mathbf{x})$ for any choice of $\alpha \neq 0$. This becomes an equality for $|\alpha| \rightarrow \infty$ (the approximate equality is due to the offset $\epsilon$ for the variance).

Another advantage of the layer normalization is that it does not depend on the minibatch size. It is also independent of whether we are in training or test regime. In other words, it is simply a deterministic transformation that standardizes the activations to a given scale. This can be very beneficial in preventing divergence in optimization. We skip further details and recommend that interested readers consult the original paper.

### Batch Normalization During Prediction

As we mentioned earlier, batch normalization typically behaves differently in training mode than in prediction mode. First, the noise in the sample mean and the sample variance arising from estimating each on minibatches is no longer desirable once we have trained the model. Second, we might not have the luxury of computing per-batch normalization statistics. For example, we might need to apply our model to make one prediction at a time.

Typically, after training, we use the entire dataset to compute stable estimates of the variable statistics and then fix them at prediction time. Hence, batch normalization behaves differently during training than at test time. Recall that dropout also exhibits this characteristic.

## 8.5.3 Implementation from Scratch

To see how batch normalization works in practice, we implement one from scratch below.

```
def batch_norm(X, gamma, beta, moving_mean, moving_var, eps, momentum):
    # Use is_grad_enabled to determine whether we are in training mode
    if not torch.is_grad_enabled():
        # In prediction mode, use mean and variance obtained by moving average
        X_hat = (X - moving_mean) / torch.sqrt(moving_var + eps)
    else:
        assert len(X.shape) in (2, 4)
        if len(X.shape) == 2:
            # When using a fully connected layer, calculate the mean and
            # variance on the feature dimension
            mean = X.mean(dim=0)
            var = ((X - mean) ** 2).mean(dim=0)
        else:
            # When using a two-dimensional convolutional layer, calculate the
            # mean and variance on the channel dimension (axis=1). Here we
            # need to maintain the shape of X, so that the broadcasting
            # operation can be carried out later
            mean = X.mean(dim=(0, 2, 3), keepdim=True)
            var = ((X - mean) ** 2).mean(dim=(0, 2, 3), keepdim=True)
        # In training mode, the current mean and variance are used
        X_hat = (X - mean) / torch.sqrt(var + eps)
        # Update the mean and variance using moving average
        moving_mean = (1.0 - momentum) * moving_mean + momentum * mean
        moving_var = (1.0 - momentum) * moving_var + momentum * var
    Y = gamma * X_hat + beta  # Scale and shift
    return Y, moving_mean.data, moving_var.data
```

We can now create a proper `BatchNorm` layer. Our layer will maintain proper parameters for scale `gamma` and shift `beta`, both of which will be updated in the course of training. Additionally, our layer will maintain moving averages of the means and variances for subsequent use during model prediction.

Putting aside the algorithmic details, note the design pattern underlying our implementation of the layer. Typically, we define the mathematics in a separate function, say `batch_norm`.

We then integrate this functionality into a custom layer, whose code mostly addresses bookkeeping matters, such as moving data to the right device context, allocating and initializing any required variables, keeping track of moving averages (here for mean and variance), and so on. This pattern enables a clean separation of mathematics from boilerplate code. Also note that for the sake of convenience we did not worry about automatically inferring the input shape here; thus we need to specify the number of features throughout. By now all modern deep learning frameworks offer automatic detection of size and shape in the high-level batch normalization APIs (in practice we will use this instead).

```
class BatchNorm(nn.Module):
    # num_features: the number of outputs for a fully connected layer or the
    # number of output channels for a convolutional layer. num_dims: 2 for a
    # fully connected layer and 4 for a convolutional layer
    def __init__(self, num_features, num_dims):
        super().__init__()
        if num_dims == 2:
            shape = (1, num_features)
        else:
            shape = (1, num_features, 1, 1)
        # The scale parameter and the shift parameter (model parameters) are
        # initialized to 1 and 0, respectively
        self.gamma = nn.Parameter(torch.ones(shape))
        self.beta = nn.Parameter(torch.zeros(shape))
        # The variables that are not model parameters are initialized to 0 and
        # 1
        self.moving_mean = torch.zeros(shape)
        self.moving_var = torch.ones(shape)

    def forward(self, X):
        # If X is not on the main memory, copy moving_mean and moving_var to
        # the device where X is located
        if self.moving_mean.device != X.device:
            self.moving_mean = self.moving_mean.to(X.device)
            self.moving_var = self.moving_var.to(X.device)
        # Save the updated moving_mean and moving_var
        Y, self.moving_mean, self.moving_var = batch_norm(
            X, self.gamma, self.beta, self.moving_mean,
            self.moving_var, eps=1e-5, momentum=0.1)
        return Y
```

We used `momentum` to govern the aggregation over past mean and variance estimates. This is somewhat of a misnomer as it has nothing whatsoever to do with the *momentum* term of optimization. Nonetheless, it is the commonly adopted name for this term and in deference to API naming convention we use the same variable name in our code.

### 8.5.4 LeNet with Batch Normalization

To see how to apply `BatchNorm` in context, below we apply it to a traditional LeNet model (Section 7.6). Recall that batch normalization is applied after the convolutional layers or fully connected layers but before the corresponding activation functions.

```
class BNLeNetScratch(d2l.Classifier):
    def __init__(self, lr=0.1, num_classes=10):
        super().__init__()
        self.save_hyperparameters()
        self.net = nn.Sequential(
            nn.LazyConv2d(6, kernel_size=5), BatchNorm(6, num_dims=4),
            nn.Sigmoid(), nn.AvgPool2d(kernel_size=2, stride=2),
            nn.LazyConv2d(16, kernel_size=5), BatchNorm(16, num_dims=4),
            nn.Sigmoid(), nn.AvgPool2d(kernel_size=2, stride=2),
            nn.Flatten(), nn.LazyLinear(120),
            BatchNorm(120, num_dims=2), nn.Sigmoid(), nn.LazyLinear(84),
            BatchNorm(84, num_dims=2), nn.Sigmoid(),
            nn.LazyLinear(num_classes))
```

As before, we will train our network on the Fashion-MNIST dataset. This code is virtually identical to that when we first trained LeNet.

```
trainer = d2l.Trainer(max_epochs=10, num_gpus=1)
data = d2l.FashionMNIST(batch_size=128)
model = BNLeNetScratch(lr=0.1)
model.apply_init([next(iter(data.get_dataloader(True)))[0]], d2l.init_cnn)
trainer.fit(model, data)
```

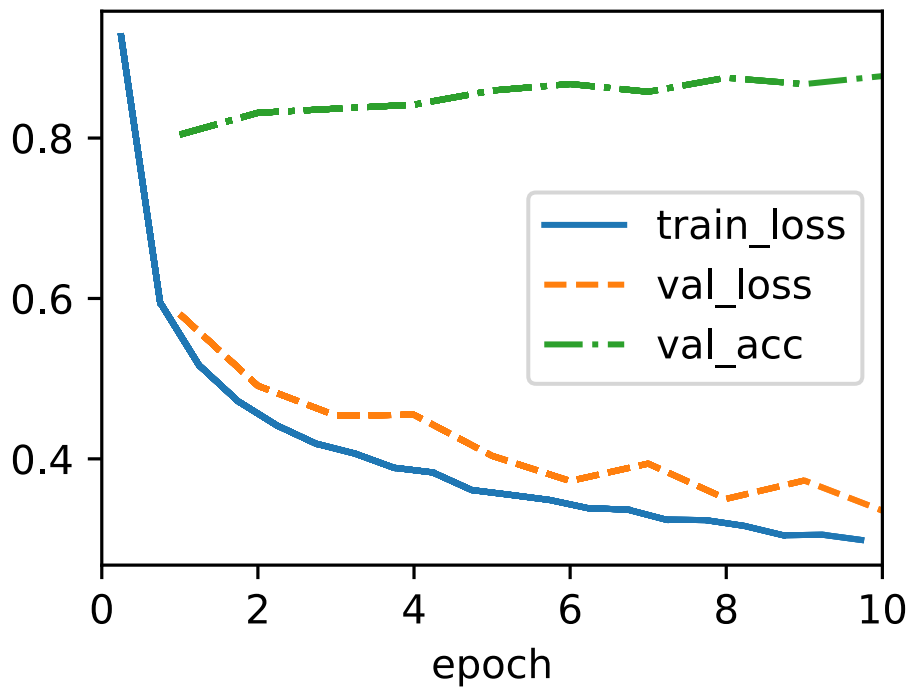

Let's have a look at the scale parameter `gamma` and the shift parameter `beta` learned from the first batch normalization layer.

```
model.net[1].gamma.reshape((-1,)), model.net[1].beta.reshape((-1,))
```

```
(tensor([2.2417, 1.8804, 2.0719, 1.1316, 1.9629, 1.6801], device='cuda:0',
        grad_fn=<ReshapeAliasBackward0>),
 tensor([ 0.3080, -1.0170,  1.4802, -0.2111, -1.9058,  0.6248], device='cuda:0
↪',
        grad_fn=<ReshapeAliasBackward0>))
```

### 8.5.5 Concise Implementation

Compared with the `BatchNorm` class, which we just defined ourselves, we can use the `BatchNorm` class defined in high-level APIs from the deep learning framework directly. The code looks virtually identical to our implementation above, except that we no longer need to provide additional arguments for it to get the dimensions right.

```
class BNLeNet(d2l.Classifier):
    def __init__(self, lr=0.1, num_classes=10):
        super().__init__()
        self.save_hyperparameters()
        self.net = nn.Sequential(
            nn.LazyConv2d(6, kernel_size=5), nn.LazyBatchNorm2d(),
            nn.Sigmoid(), nn.AvgPool2d(kernel_size=2, stride=2),
            nn.LazyConv2d(16, kernel_size=5), nn.LazyBatchNorm2d(),
            nn.Sigmoid(), nn.AvgPool2d(kernel_size=2, stride=2),
            nn.Flatten(), nn.LazyLinear(120), nn.LazyBatchNorm1d(),
            nn.Sigmoid(), nn.LazyLinear(84), nn.LazyBatchNorm1d(),
            nn.Sigmoid(), nn.LazyLinear(num_classes))
```

Below, we use the same hyperparameters to train our model. Note that as usual, the high-level API variant runs much faster because its code has been compiled to C++ or CUDA while our custom implementation must be interpreted by Python.

```
trainer = d2l.Trainer(max_epochs=10, num_gpus=1)
data = d2l.FashionMNIST(batch_size=128)
model = BNLeNet(lr=0.1)
model.apply_init([next(iter(data.get_dataloader(True)))[0]], d2l.init_cnn)
trainer.fit(model, data)
```

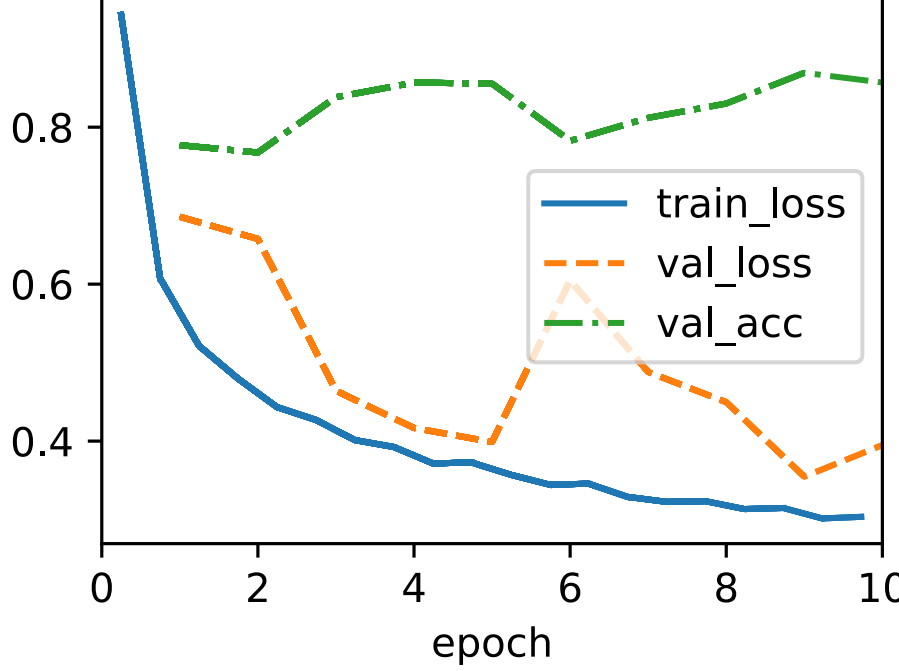

### 8.5.6 Discussion

Intuitively, batch normalization is thought to make the optimization landscape smoother. However, we must be careful to distinguish between speculative intuitions and true explanations for the phenomena that we observe when training deep models. Recall that we do not even know why simpler deep neural networks (MLPs and conventional CNNs) generalize well in the first place. Even with dropout and weight decay, they remain so flexible that their ability to generalize to unseen data likely needs significantly more refined learning-theoretic generalization guarantees.

The original paper proposing batch normalization (Ioffe and Szegedy, 2015), in addition to introducing a powerful and useful tool, offered an explanation for why it works: by reducing *internal covariate shift*. Presumably by *internal covariate shift* they meant something like the intuition expressed above—the notion that the distribution of variable values changes over the course of training. However, there were two problems with this explanation: i) This drift is very different from *covariate shift*, rendering the name a misnomer. If anything, it is closer to concept drift. ii) The explanation offers an under-specified intuition but leaves the question of *why precisely this technique works* an open question wanting for a rigorous explanation. Throughout this book, we aim to convey the intuitions that practitioners use to guide their development of deep neural networks. However, we believe that it is important to separate these guiding intuitions from established scientific fact. Eventually, when you master this material and start writing your own research papers you will want to be clear to delineate between technical claims and hunches.

Following the success of batch normalization, its explanation in terms of *internal covariate shift* has repeatedly surfaced in debates in the technical literature and broader discourse about how to present machine learning research. In a memorable speech given while accepting a Test of Time Award at the 2017 NeurIPS conference, Ali Rahimi used *internal covariate shift* as a focal point in an argument likening the modern practice of deep learning to alchemy. Subsequently, the example was revisited in detail in a position paper outlining troubling trends in machine learning (Lipton and Steinhardt, 2018). Other authors have proposed alternative explanations for the success of batch normalization, some (Santurkar *et al.*, 2018) claiming that batch normalization's success comes despite exhibiting behavior that is in some ways opposite to those claimed in the original paper.

We note that the *internal covariate shift* is no more worthy of criticism than any of thousands of similarly vague claims made every year in the technical machine learning literature. Likely, its resonance as a focal point of these debates owes to its broad recognizability for the target audience. Batch normalization has proven an indispensable method, applied in nearly all deployed image classifiers, earning the paper that introduced the technique tens of thousands of citations. We conjecture, though, that the guiding principles of regularization through noise injection, acceleration through rescaling and lastly preprocessing may well lead to further inventions of layers and techniques in the future.

On a more practical note, there are a number of aspects worth remembering about batch normalization:

- During model training, batch normalization continuously adjusts the intermediate output of the network by utilizing the mean and standard deviation of the minibatch, so that the values of the intermediate output in each layer throughout the neural network are more stable.

- Batch normalization is slightly different for fully connected layers than for convolutional

layers. In fact, for convolutional layers, layer normalization can sometimes be used as an alternative.

- Like a dropout layer, batch normalization layers have different behaviors in training mode than in prediction mode.
- Batch normalization is useful for regularization and improving convergence in optimization. By contrast, the original motivation of reducing internal covariate shift seems not to be a valid explanation.
- For more robust models that are less sensitive to input perturbations, consider removing batch normalization (Wang *et al.*, 2022).

### 8.5.7 Exercises

1. Should we remove the bias parameter from the fully connected layer or the convolutional layer before the batch normalization? Why?
2. Compare the learning rates for LeNet with and without batch normalization.
   1. Plot the increase in validation accuracy.
   2. How large can you make the learning rate before the optimization fails in both cases?
3. Do we need batch normalization in every layer? Experiment with it.
4. Implement a "lite" version of batch normalization that only removes the mean, or alternatively one that only removes the variance. How does it behave?
5. Fix the parameters `beta` and `gamma`. Observe and analyze the results.
6. Can you replace dropout by batch normalization? How does the behavior change?
7. Research ideas: think of other normalization transforms that you can apply:
   1. Can you apply the probability integral transform?
   2. Can you use a full-rank covariance estimate? Why should you probably not do that?
   3. Can you use other compact matrix variants (block-diagonal, low-displacement rank, Monarch, etc.)?
   4. Does a sparsification compression act as a regularizer?
   5. Are there other projections (e.g., convex cone, symmetry group-specific transforms) that you can use?

Discussions[132].

# 8.6 Residual Networks (ResNet) and ResNeXt

As we design ever deeper networks it becomes imperative to understand how adding layers can increase the complexity and expressiveness of the network. Even more important is the ability to design networks where adding layers makes networks strictly more expressive rather than just different. To make some progress we need a bit of mathematics.

```
import torch
from torch import nn
from torch.nn import functional as F
from d2l import torch as d2l
```

## 8.6.1 Function Classes

Consider $\mathcal{F}$, the class of functions that a specific network architecture (together with learning rates and other hyperparameter settings) can reach. That is, for all $f \in \mathcal{F}$ there exists some set of parameters (e.g., weights and biases) that can be obtained through training on a suitable dataset. Let's assume that $f^*$ is the "truth" function that we really would like to find. If it is in $\mathcal{F}$, we are in good shape but typically we will not be quite so lucky. Instead, we will try to find some $f^*_\mathcal{F}$ which is our best bet within $\mathcal{F}$. For instance, given a dataset with features $\mathbf{X}$ and labels $\mathbf{y}$, we might try finding it by solving the following optimization problem:

$$f^*_\mathcal{F} \stackrel{\textrm{def}}{=} \mathop{\mathrm{argmin}}_f L(\mathbf{X}, \mathbf{y}, f) \textrm{ subject to } f \in \mathcal{F}. \tag{8.6.1}$$

We know that regularization (Morozov, 1984, Tikhonov and Arsenin, 1977) may control complexity of $\mathcal{F}$ and achieve consistency, so a larger size of training data generally leads to better $f^*_\mathcal{F}$. It is only reasonable to assume that if we design a different and more powerful architecture $\mathcal{F}'$ we should arrive at a better outcome. In other words, we would expect that $f^*_{\mathcal{F}'}$ is "better" than $f^*_\mathcal{F}$. However, if $\mathcal{F} \not\subseteq \mathcal{F}'$ there is no guarantee that this should even happen. In fact, $f^*_{\mathcal{F}'}$ might well be worse. As illustrated by Fig. 8.6.1, for non-nested function classes, a larger function class does not always move closer to the "truth" function $f^*$. For instance, on the left of Fig. 8.6.1, though $\mathcal{F}_3$ is closer to $f^*$ than $\mathcal{F}_1$, $\mathcal{F}_6$ moves away and there is no guarantee that further increasing the complexity can reduce the distance from $f^*$. With nested function classes where $\mathcal{F}_1 \subseteq \cdots \subseteq \mathcal{F}_6$ on the right of Fig. 8.6.1, we can avoid the aforementioned issue from the non-nested function classes.

Thus, only if larger function classes contain the smaller ones are we guaranteed that increasing them strictly increases the expressive power of the network. For deep neural networks, if we can train the newly-added layer into an identity function $f(\mathbf{x}) = \mathbf{x}$, the new model will be as effective as the original model. As the new model may get a better solution to fit the training dataset, the added layer might make it easier to reduce training errors.

This is the question that He *et al.* (2016) considered when working on very deep computer vision models. At the heart of their proposed *residual network* (*ResNet*) is the idea that every additional layer should more easily contain the identity function as one of its elements. These considerations are rather profound but they led to a surprisingly simple solution, a *residual block*. With it, ResNet won the ImageNet Large Scale Visual Recognition Challenge in 2015. The design had a profound influence on how to build deep neural networks. For instance, residual blocks have been added to recurrent networks (Kim *et al.*, 2017, Prakash *et al.*, 2016). Likewise, Transformers (Vaswani *et al.*, 2017) use them to stack many layers of networks efficiently. It is also used in graph neural networks (Kipf and Welling, 2016) and, as a basic concept, it has been used extensively in computer vision (Redmon and Farhadi, 2018, Ren *et al.*, 2015). Note that residual networks are predated by highway networks (Srivastava *et al.*, 2015) that share some of the motivation, albeit without the elegant parametrization around the identity function.

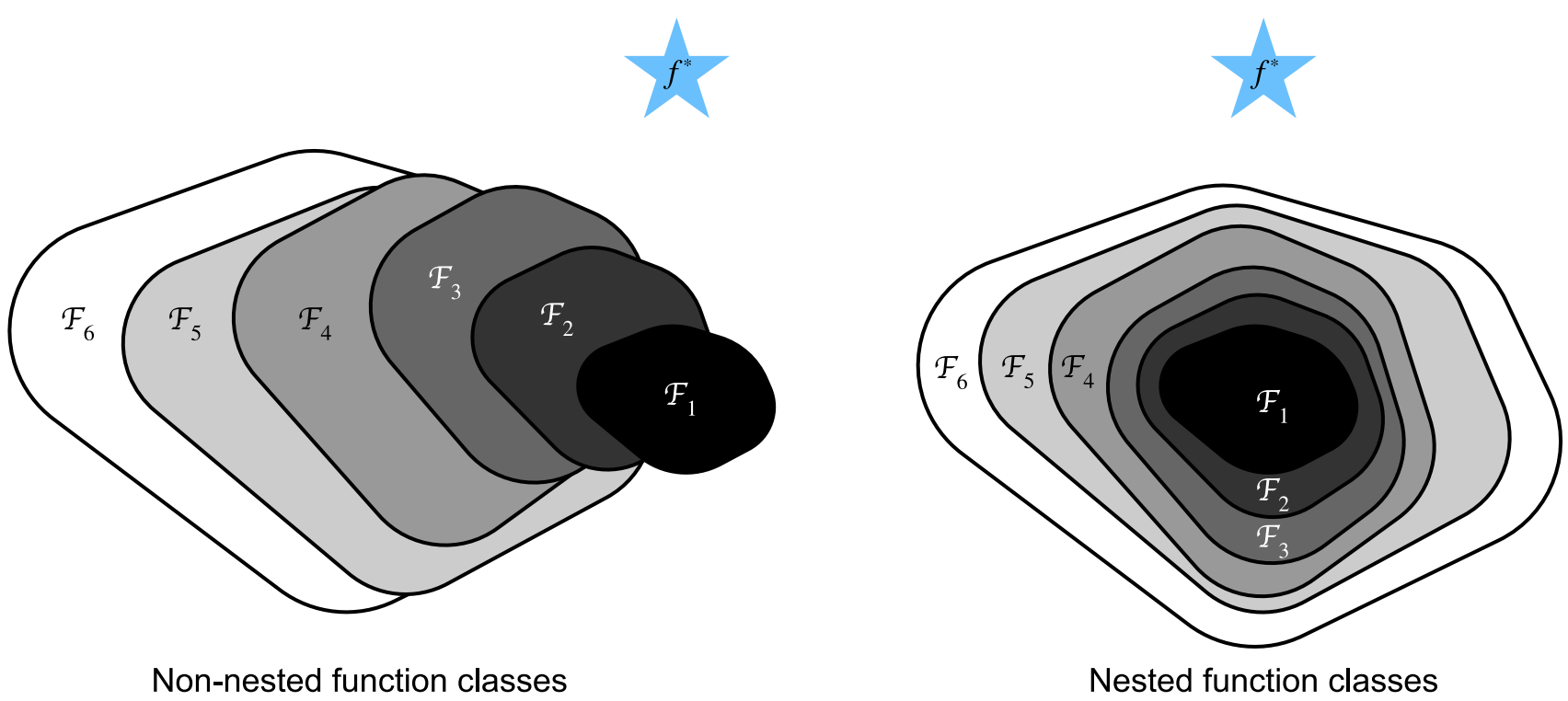

Fig. 8.6.1 For non-nested function classes, a larger (indicated by area) function class does not guarantee we will get closer to the "truth" function ($f^*$). This does not happen in nested function classes.

## 8.6.2 Residual Blocks

Let's focus on a local part of a neural network, as depicted in Fig. 8.6.2. Denote the input by $\mathbf{x}$. We assume that $f(\mathbf{x})$, the desired underlying mapping we want to obtain by learning, is to be used as input to the activation function on the top. On the left, the portion within the dotted-line box must directly learn $f(\mathbf{x})$. On the right, the portion within the dotted-line box needs to learn the *residual mapping* $g(\mathbf{x}) = f(\mathbf{x}) - \mathbf{x}$, which is how the residual block derives its name. If the identity mapping $f(\mathbf{x}) = \mathbf{x}$ is the desired underlying mapping, the residual mapping amounts to $g(\mathbf{x}) = 0$ and it is thus easier to learn: we only need to push the weights and biases of the upper weight layer (e.g., fully connected layer and convolutional layer) within the dotted-line box to zero. The right figure illustrates the *residual block* of ResNet, where the solid line carrying the layer input $\mathbf{x}$ to the addition operator is called a *residual connection* (or *shortcut connection*). With residual blocks, inputs can forward propagate faster through the residual connections across layers. In fact, the residual block can be thought of as a special case of the multi-branch Inception block: it has two branches one of which is the identity mapping.

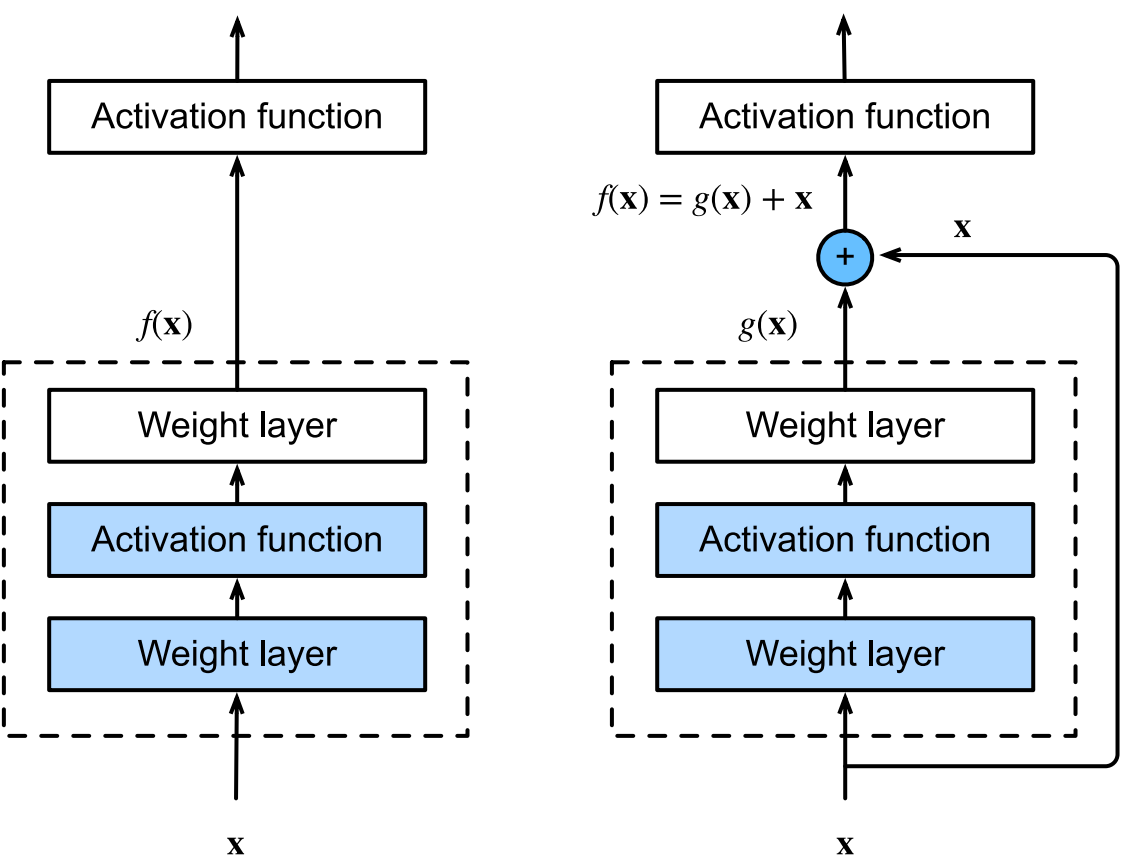

Fig. 8.6.2 In a regular block (left), the portion within the dotted-line box must directly learn the mapping $f(\mathbf{x})$. In a residual block (right), the portion within the dotted-line box needs to learn the residual mapping $g(\mathbf{x}) = f(\mathbf{x}) - \mathbf{x}$, making the identity mapping $f(\mathbf{x}) = \mathbf{x}$ easier to learn.

ResNet has VGG's full $3 \times 3$ convolutional layer design. The residual block has two $3 \times 3$ convolutional layers with the same number of output channels. Each convolutional layer is followed by a batch normalization layer and a ReLU activation function. Then, we skip these two convolution operations and add the input directly before the final ReLU activation function. This kind of design requires that the output of the two convolutional layers has to be of the same shape as the input, so that they can be added together. If we want to change the number of channels, we need to introduce an additional $1 \times 1$ convolutional layer to transform the input into the desired shape for the addition operation. Let's have a look at the code below.

```
class Residual(nn.Module):  #@save
    """The Residual block of ResNet models."""
    def __init__(self, num_channels, use_1x1conv=False, strides=1):
        super().__init__()
        self.conv1 = nn.LazyConv2d(num_channels, kernel_size=3, padding=1,
                                   stride=strides)
        self.conv2 = nn.LazyConv2d(num_channels, kernel_size=3, padding=1)
        if use_1x1conv:
            self.conv3 = nn.LazyConv2d(num_channels, kernel_size=1,
                                       stride=strides)
        else:
            self.conv3 = None
        self.bn1 = nn.LazyBatchNorm2d()
        self.bn2 = nn.LazyBatchNorm2d()

    def forward(self, X):
        Y = F.relu(self.bn1(self.conv1(X)))
```

```
        Y = self.bn2(self.conv2(Y))
        if self.conv3:
            X = self.conv3(X)
        Y += X
        return F.relu(Y)
```

This code generates two types of networks: one where we add the input to the output before applying the ReLU nonlinearity whenever `use_1x1conv=False`; and one where we adjust channels and resolution by means of a $1 \times 1$ convolution before adding. Fig. 8.6.3 illustrates this.

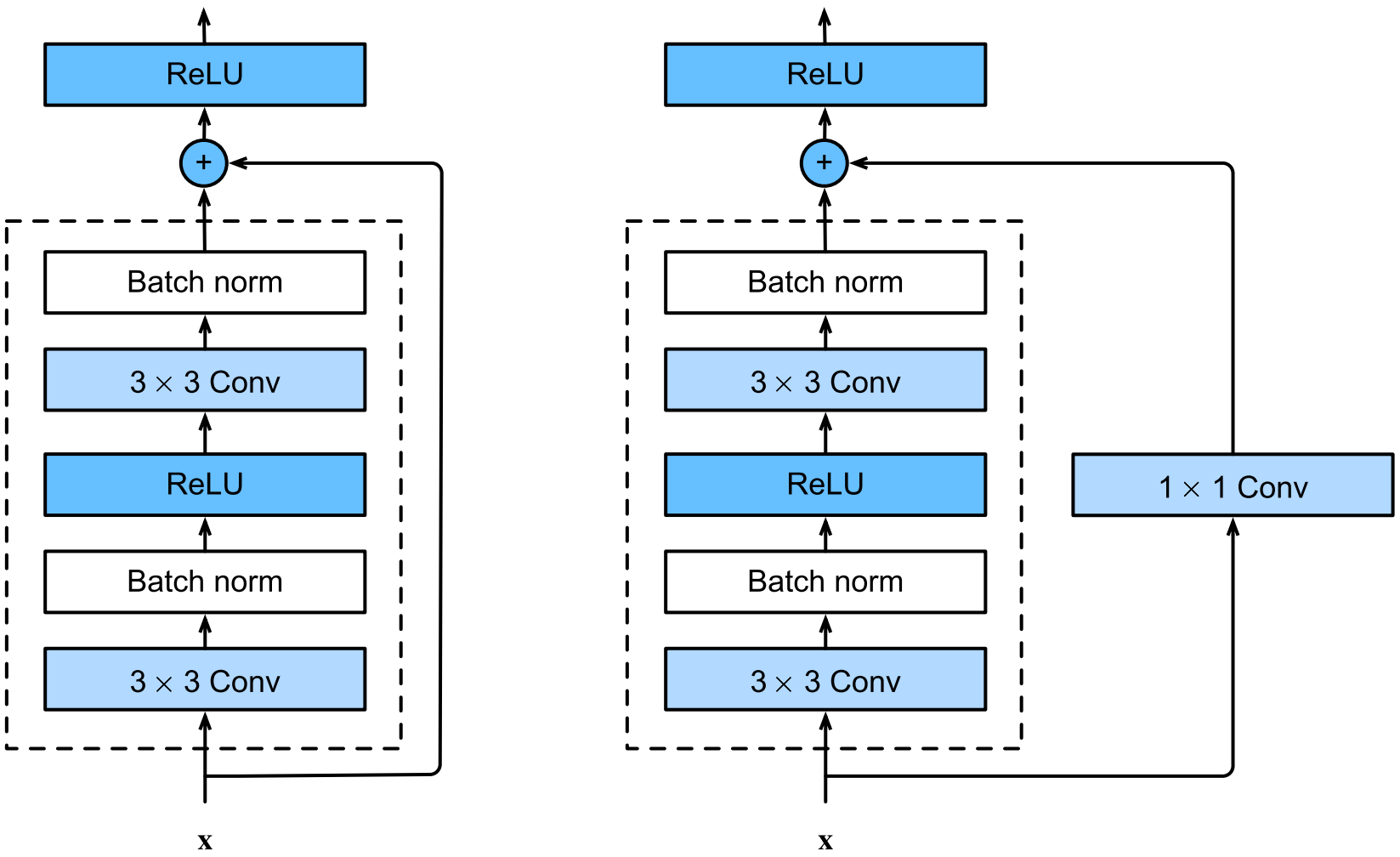

Fig. 8.6.3 ResNet block with and without $1 \times 1$ convolution, which transforms the input into the desired shape for the addition operation.

Now let's look at a situation where the input and output are of the same shape, where $1 \times 1$ convolution is not needed.

```
blk = Residual(3)
X = torch.randn(4, 3, 6, 6)
blk(X).shape
```

```
torch.Size([4, 3, 6, 6])
```

We also have the option to halve the output height and width while increasing the number of output channels. In this case we use $1 \times 1$ convolutions via `use_1x1conv=True`. This comes in handy at the beginning of each ResNet block to reduce the spatial dimensionality via `strides=2`.

```
blk = Residual(6, use_1x1conv=True, strides=2)
blk(X).shape
```

```
torch.Size([4, 6, 3, 3])
```

### 8.6.3 ResNet Model

The first two layers of ResNet are the same as those of the GoogLeNet we described before: the $7 \times 7$ convolutional layer with 64 output channels and a stride of 2 is followed by the $3 \times 3$ max-pooling layer with a stride of 2. The difference is the batch normalization layer added after each convolutional layer in ResNet.

```
class ResNet(d2l.Classifier):
    def b1(self):
        return nn.Sequential(
            nn.LazyConv2d(64, kernel_size=7, stride=2, padding=3),
            nn.LazyBatchNorm2d(), nn.ReLU(),
            nn.MaxPool2d(kernel_size=3, stride=2, padding=1))
```

GoogLeNet uses four modules made up of Inception blocks. However, ResNet uses four modules made up of residual blocks, each of which uses several residual blocks with the same number of output channels. The number of channels in the first module is the same as the number of input channels. Since a max-pooling layer with a stride of 2 has already been used, it is not necessary to reduce the height and width. In the first residual block for each of the subsequent modules, the number of channels is doubled compared with that of the previous module, and the height and width are halved.

```
@d2l.add_to_class(ResNet)
def block(self, num_residuals, num_channels, first_block=False):
    blk = []
    for i in range(num_residuals):
        if i == 0 and not first_block:
            blk.append(Residual(num_channels, use_1x1conv=True, strides=2))
        else:
            blk.append(Residual(num_channels))
    return nn.Sequential(*blk)
```

Then, we add all the modules to ResNet. Here, two residual blocks are used for each module. Lastly, just like GoogLeNet, we add a global average pooling layer, followed by the fully connected layer output.

```
@d2l.add_to_class(ResNet)
def __init__(self, arch, lr=0.1, num_classes=10):
    super(ResNet, self).__init__()
    self.save_hyperparameters()
    self.net = nn.Sequential(self.b1())
```

```
    for i, b in enumerate(arch):
        self.net.add_module(f'b{i+2}', self.block(*b, first_block=(i==0)))
    self.net.add_module('last', nn.Sequential(
        nn.AdaptiveAvgPool2d((1, 1)), nn.Flatten(),
        nn.LazyLinear(num_classes)))
    self.net.apply(d2l.init_cnn)
```

There are four convolutional layers in each module (excluding the $1 \times 1$ convolutional layer). Together with the first $7 \times 7$ convolutional layer and the final fully connected layer, there are 18 layers in total. Therefore, this model is commonly known as ResNet-18. By configuring different numbers of channels and residual blocks in the module, we can create different ResNet models, such as the deeper 152-layer ResNet-152. Although the main architecture of ResNet is similar to that of GoogLeNet, ResNet's structure is simpler and easier to modify. All these factors have resulted in the rapid and widespread use of ResNet. Fig. 8.6.4 depicts the full ResNet-18.

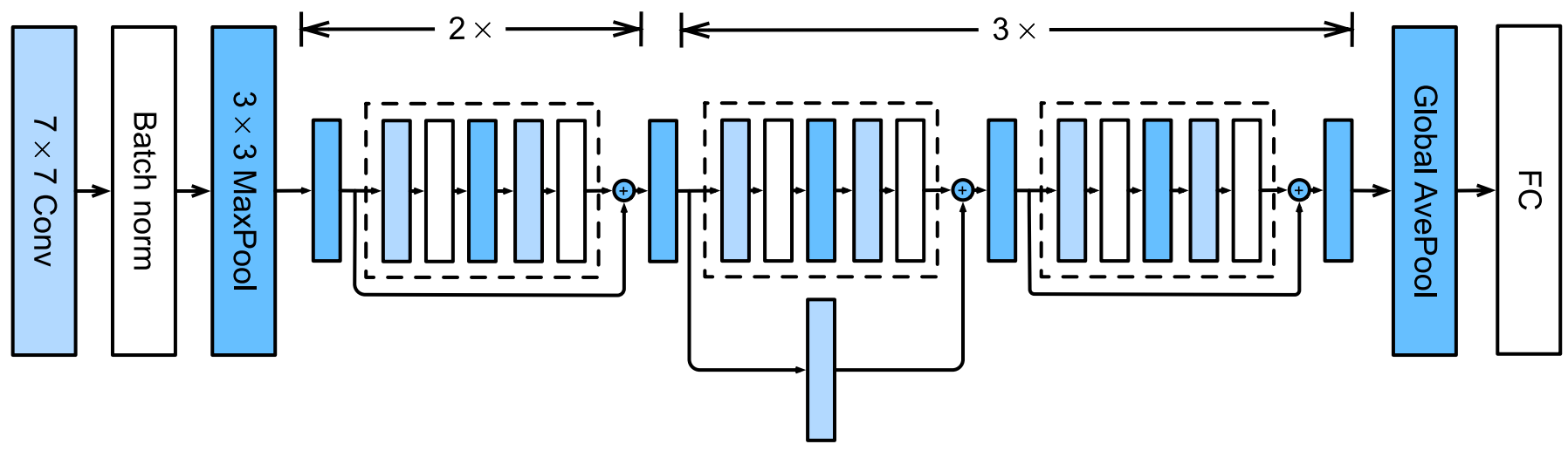

Fig. 8.6.4 The ResNet-18 architecture.

Before training ResNet, let's observe how the input shape changes across different modules in ResNet. As in all the previous architectures, the resolution decreases while the number of channels increases up until the point where a global average pooling layer aggregates all features.

```
class ResNet18(ResNet):
    def __init__(self, lr=0.1, num_classes=10):
        super().__init__(((2, 64), (2, 128), (2, 256), (2, 512)),
                       lr, num_classes)
```

```
ResNet18().layer_summary((1, 1, 96, 96))
```

```
Sequential output shape:     torch.Size([1, 64, 24, 24])
Sequential output shape:     torch.Size([1, 64, 24, 24])
Sequential output shape:     torch.Size([1, 128, 12, 12])
Sequential output shape:     torch.Size([1, 256, 6, 6])
Sequential output shape:     torch.Size([1, 512, 3, 3])
Sequential output shape:     torch.Size([1, 10])
```

## 8.6.4 Training

We train ResNet on the Fashion-MNIST dataset, just like before. ResNet is quite a powerful and flexible architecture. The plot capturing training and validation loss illustrates a significant gap between both graphs, with the training loss being considerably lower. For a network of this flexibility, more training data would offer distinct benefit in closing the gap and improving accuracy.

```
model = ResNet18(lr=0.01)
trainer = d2l.Trainer(max_epochs=10, num_gpus=1)
data = d2l.FashionMNIST(batch_size=128, resize=(96, 96))
model.apply_init([next(iter(data.get_dataloader(True)))[0]], d2l.init_cnn)
trainer.fit(model, data)
```

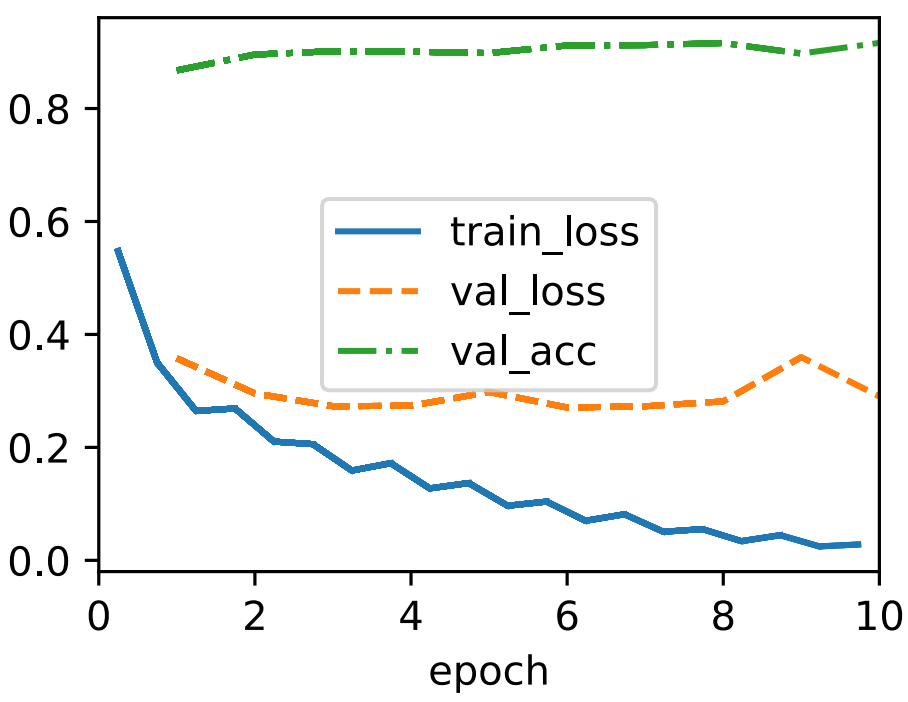

## 8.6.5 ResNeXt

One of the challenges one encounters in the design of ResNet is the trade-off between non-linearity and dimensionality within a given block. That is, we could add more nonlinearity by increasing the number of layers, or by increasing the width of the convolutions. An alternative strategy is to increase the number of channels that can carry information between blocks. Unfortunately, the latter comes with a quadratic penalty since the computational cost of ingesting $c_\textrm{i}$ channels and emitting $c_\textrm{o}$ channels is proportional to $\mathcal{O}(c_\textrm{i} \cdot c_\textrm{o})$ (see our discussion in Section 7.4).

We can take some inspiration from the Inception block of Fig. 8.4.1 which has information flowing through the block in separate groups. Applying the idea of multiple independent groups to the ResNet block of Fig. 8.6.3 led to the design of ResNeXt (Xie *et al.*, 2017). Different from the smorgasbord of transformations in Inception, ResNeXt adopts the *same* transformation in all branches, thus minimizing the need for manual tuning of each branch.

Breaking up a convolution from $c_\textrm{i}$ to $c_\textrm{o}$ channels into one of $g$ groups of size $c_\textrm{i}/g$ generating $g$ outputs of size $c_\textrm{o}/g$ is called, quite fittingly, a *grouped convolution*. The computational cost (proportionally) is reduced from $\mathcal{O}(c_\textrm{i} \cdot c_\textrm{o})$ to $\mathcal{O}(g \cdot (c_\textrm{i}/g) \cdot (c_\textrm{o}/g)) = \mathcal{O}(c_\textrm{i} \cdot c_\textrm{o} / g)$, i.e., it is $g$ times faster. Even better, the number of parameters needed to generate the output is also reduced from a $c_\textrm{i} \times c_\textrm{o}$ matrix to $g$ smaller matrices of size $(c_\textrm{i}/g) \times (c_\textrm{o}/g)$, again a $g$ times reduction. In what follows we assume that both $c_\textrm{i}$ and $c_\textrm{o}$ are divisible by $g$.

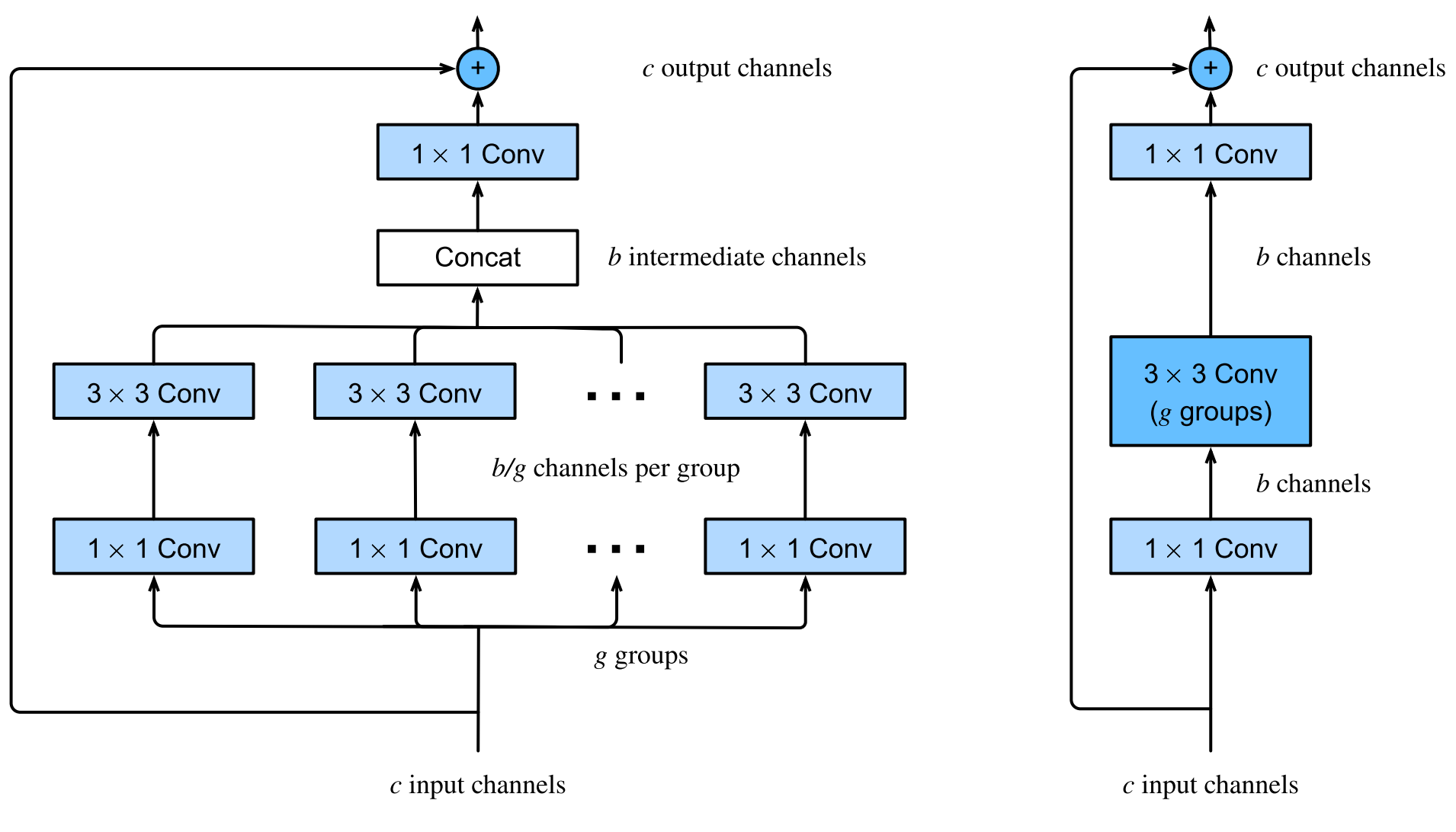

Fig. 8.6.5 The ResNeXt block. The use of grouped convolution with $g$ groups is $g$ times faster than a dense convolution. It is a bottleneck residual block when the number of intermediate channels $b$ is less than $c$.

The only challenge in this design is that no information is exchanged between the $g$ groups. The ResNeXt block of Fig. 8.6.5 amends this in two ways: the grouped convolution with a $3 \times 3$ kernel is sandwiched in between two $1 \times 1$ convolutions. The second one serves double duty in changing the number of channels back. The benefit is that we only pay the $\mathcal{O}(c \cdot b)$ cost for $1 \times 1$ kernels and can make do with an $\mathcal{O}(b^2 / g)$ cost for $3 \times 3$ kernels. Similar to the residual block implementation in Section 8.6.2, the residual connection is replaced (thus generalized) by a $1 \times 1$ convolution.

The right-hand figure in Fig. 8.6.5 provides a much more concise summary of the resulting network block. It will also play a major role in the design of generic modern CNNs in Section 8.8. Note that the idea of grouped convolutions dates back to the implementation of AlexNet (Krizhevsky *et al.*, 2012). When distributing the network across two GPUs with limited memory, the implementation treated each GPU as its own channel with no ill effects.

The following implementation of the `ResNeXtBlock` class takes as argument `groups` ($g$), with `bot_channels` ($b$) intermediate (bottleneck) channels. Lastly, when we need to reduce the height and width of the representation, we add a stride of 2 by setting `use_1x1conv=True, strides=2`.

```
class ResNeXtBlock(nn.Module):  #@save
```

```
    """The ResNeXt block."""
    def __init__(self, num_channels, groups, bot_mul, use_1x1conv=False,
                 strides=1):
        super().__init__()
        bot_channels = int(round(num_channels * bot_mul))
        self.conv1 = nn.LazyConv2d(bot_channels, kernel_size=1, stride=1)
        self.conv2 = nn.LazyConv2d(bot_channels, kernel_size=3,
                                   stride=strides, padding=1,
                                   groups=bot_channels//groups)
        self.conv3 = nn.LazyConv2d(num_channels, kernel_size=1, stride=1)
        self.bn1 = nn.LazyBatchNorm2d()
        self.bn2 = nn.LazyBatchNorm2d()
        self.bn3 = nn.LazyBatchNorm2d()
        if use_1x1conv:
            self.conv4 = nn.LazyConv2d(num_channels, kernel_size=1,
                                       stride=strides)
            self.bn4 = nn.LazyBatchNorm2d()
        else:
            self.conv4 = None

    def forward(self, X):
        Y = F.relu(self.bn1(self.conv1(X)))
        Y = F.relu(self.bn2(self.conv2(Y)))
        Y = self.bn3(self.conv3(Y))
        if self.conv4:
            X = self.bn4(self.conv4(X))
        return F.relu(Y + X)
```

Its use is entirely analogous to that of the `ResNetBlock` discussed previously. For instance, when using (`use_1x1conv=False, strides=1`), the input and output are of the same shape. Alternatively, setting `use_1x1conv=True, strides=2` halves the output height and width.

```
blk = ResNeXtBlock(32, 16, 1)
X = torch.randn(4, 32, 96, 96)
blk(X).shape
```

```
torch.Size([4, 32, 96, 96])
```

## 8.6.6 Summary and Discussion

Nested function classes are desirable since they allow us to obtain strictly *more powerful* rather than also subtly *different* function classes when adding capacity. One way of accomplishing this is by letting additional layers to simply pass through the input to the output. Residual connections allow for this. As a consequence, this changes the inductive bias from simple functions being of the form $f(\mathbf{x}) = 0$ to simple functions looking like $f(\mathbf{x}) = \mathbf{x}$.

The residual mapping can learn the identity function more easily, such as pushing parameters in the weight layer to zero. We can train an effective *deep* neural network by having residual blocks. Inputs can forward propagate faster through the residual connections across layers. As a consequence, we can thus train much deeper networks. For instance, the original ResNet paper (He *et al.*, 2016) allowed for up to 152 layers. Another benefit of residual networks is that it allows us to add layers, initialized as the identity function, *during* the training process. After all, the default behavior of a layer is to let the data pass through unchanged. This can accelerate the training of very large networks in some cases.

Prior to residual connections, bypassing paths with gating units were introduced to effectively train highway networks with over 100 layers (Srivastava *et al.*, 2015). Using identity functions as bypassing paths, ResNet performed remarkably well on multiple computer vision tasks. Residual connections had a major influence on the design of subsequent deep neural networks, of either convolutional or sequential nature. As we will introduce later, the Transformer architecture (Vaswani *et al.*, 2017) adopts residual connections (together with other design choices) and is pervasive in areas as diverse as language, vision, speech, and reinforcement learning.

ResNeXt is an example for how the design of convolutional neural networks has evolved over time: by being more frugal with computation and trading it off against the size of the activations (number of channels), it allows for faster and more accurate networks at lower cost. An alternative way of viewing grouped convolutions is to think of a block-diagonal matrix for the convolutional weights. Note that there are quite a few such "tricks" that lead to more efficient networks. For instance, ShiftNet (Wu *et al.*, 2018) mimicks the effects of a $3 \times 3$ convolution, simply by adding shifted activations to the channels, offering increased function complexity, this time without any computational cost.

A common feature of the designs we have discussed so far is that the network design is fairly manual, primarily relying on the ingenuity of the designer to find the "right" network hyperparameters. While clearly feasible, it is also very costly in terms of human time and there is no guarantee that the outcome is optimal in any sense. In Section 8.8 we will discuss a number of strategies for obtaining high quality networks in a more automated fashion. In particular, we will review the notion of *network design spaces* that led to the RegNetX/Y models (Radosavovic *et al.*, 2020).

## 8.6.7 Exercises

1. What are the major differences between the Inception block in Fig. 8.4.1 and the residual block? How do they compare in terms of computation, accuracy, and the classes of functions they can describe?
2. Refer to Table 1 in the ResNet paper (He *et al.*, 2016) to implement different variants of the network.
3. For deeper networks, ResNet introduces a "bottleneck" architecture to reduce model complexity. Try to implement it.
4. In subsequent versions of ResNet, the authors changed the "convolution, batch normal-

ization, and activation" structure to the "batch normalization, activation, and convolution" structure. Make this improvement yourself. See Figure 1 in He *et al.* (2016) for details.

5. Why can't we just increase the complexity of functions without bound, even if the function classes are nested?

Discussions[133].

133

# 8.7 Densely Connected Networks (DenseNet)

ResNet significantly changed the view of how to parametrize the functions in deep networks. *DenseNet* (dense convolutional network) is to some extent the logical extension of this (Huang *et al.*, 2017). DenseNet is characterized by both the connectivity pattern where each layer connects to all the preceding layers and the concatenation operation (rather than the addition operator in ResNet) to preserve and reuse features from earlier layers. To understand how to arrive at it, let's take a small detour to mathematics.

```
import torch
from torch import nn
from d2l import torch as d2l
```

## 8.7.1 From ResNet to DenseNet

Recall the Taylor expansion for functions. At the point $x = 0$ it can be written as

$$f(x) = f(0) + x \cdot \left[ f'(0) + x \cdot \left[ \frac{f''(0)}{2!} + x \cdot \left[ \frac{f'''(0)}{3!} + \cdots \right] \right] \right]. \tag{8.7.1}$$

The key point is that it decomposes a function into terms of increasingly higher order. In a similar vein, ResNet decomposes functions into

$$f(\mathbf{x}) = \mathbf{x} + g(\mathbf{x}). \tag{8.7.2}$$

That is, ResNet decomposes $f$ into a simple linear term and a more complex nonlinear one. What if we wanted to capture (not necessarily add) information beyond two terms? One such solution is DenseNet (Huang *et al.*, 2017).

As shown in Fig. 8.7.1, the key difference between ResNet and DenseNet is that in the latter case outputs are *concatenated* (denoted by $[,]$) rather than added. As a result, we perform a mapping from $\mathbf{x}$ to its values after applying an increasingly complex sequence of functions:

$$\mathbf{x} \to \left[ \mathbf{x}, f_1(\mathbf{x}), f_2\left( \left[ \mathbf{x}, f_1\left(\mathbf{x}\right) \right] \right), f_3\left( \left[ \mathbf{x}, f_1\left(\mathbf{x}\right), f_2\left( \left[ \mathbf{x}, f_1\left(\mathbf{x}\right) \right] \right) \right] \right), \ldots \right]. \tag{8.7.3}$$

In the end, all these functions are combined in MLP to reduce the number of features again.

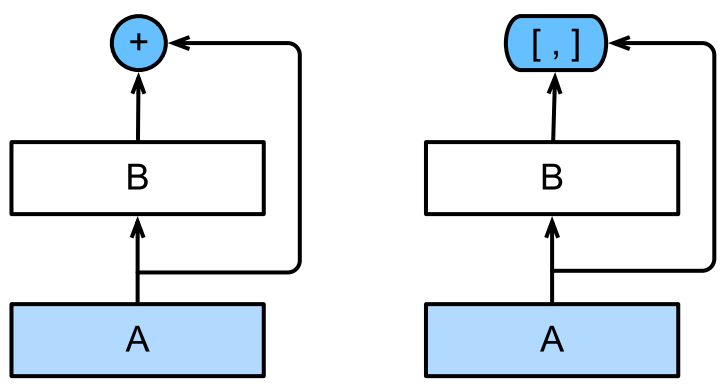

Fig. 8.7.1 The main difference between ResNet (left) and DenseNet (right) in cross-layer connections: use of addition and use of concatenation.

In terms of implementation this is quite simple: rather than adding terms, we concatenate them. The name DenseNet arises from the fact that the dependency graph between variables becomes quite dense. The final layer of such a chain is densely connected to all previous layers. The dense connections are shown in Fig. 8.7.2.

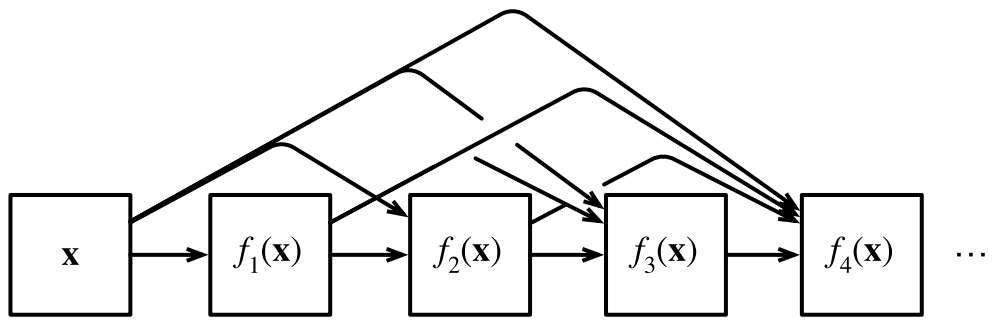

Fig. 8.7.2 Dense connections in DenseNet. Note how the dimensionality increases with depth.

The main components that comprise a DenseNet are *dense blocks* and *transition layers*. The former define how the inputs and outputs are concatenated, while the latter control the number of channels so that it is not too large, since the expansion $\mathbf{x} \to [\mathbf{x}, f_1(\mathbf{x}), f_2([\mathbf{x}, f_1(\mathbf{x})]), \ldots]$ can be quite high-dimensional.

## 8.7.2 Dense Blocks

DenseNet uses the modified "batch normalization, activation, and convolution" structure of ResNet (see the exercise in Section 8.6). First, we implement this convolution block structure.

```
def conv_block(num_channels):
    return nn.Sequential(
        nn.LazyBatchNorm2d(), nn.ReLU(),
        nn.LazyConv2d(num_channels, kernel_size=3, padding=1))
```

A *dense block* consists of multiple convolution blocks, each using the same number of output channels. In the forward propagation, however, we concatenate the input and output of each convolution block on the channel dimension. Lazy evaluation allows us to adjust the dimensionality automatically.

```
class DenseBlock(nn.Module):
    def __init__(self, num_convs, num_channels):
```

```
        super(DenseBlock, self).__init__()
        layer = []
        for i in range(num_convs):
            layer.append(conv_block(num_channels))
        self.net = nn.Sequential(*layer)

    def forward(self, X):
        for blk in self.net:
            Y = blk(X)
            # Concatenate input and output of each block along the channels
            X = torch.cat((X, Y), dim=1)
        return X
```

In the following example, we define a `DenseBlock` instance with two convolution blocks of 10 output channels. When using an input with three channels, we will get an output with $3+10+10=23$ channels. The number of convolution block channels controls the growth in the number of output channels relative to the number of input channels. This is also referred to as the *growth rate*.

```
blk = DenseBlock(2, 10)
X = torch.randn(4, 3, 8, 8)
Y = blk(X)
Y.shape
```

```
torch.Size([4, 23, 8, 8])
```

### 8.7.3 Transition Layers

Since each dense block will increase the number of channels, adding too many of them will lead to an excessively complex model. A *transition layer* is used to control the complexity of the model. It reduces the number of channels by using a $1 \times 1$ convolution. Moreover, it halves the height and width via average pooling with a stride of 2.

```
def transition_block(num_channels):
    return nn.Sequential(
        nn.LazyBatchNorm2d(), nn.ReLU(),
        nn.LazyConv2d(num_channels, kernel_size=1),
        nn.AvgPool2d(kernel_size=2, stride=2))
```

Apply a transition layer with 10 channels to the output of the dense block in the previous example. This reduces the number of output channels to 10, and halves the height and width.

```
blk = transition_block(10)
blk(Y).shape
```

```
torch.Size([4, 10, 4, 4])
```

### 8.7.4 DenseNet Model

Next, we will construct a DenseNet model. DenseNet first uses the same single convolutional layer and max-pooling layer as in ResNet.

```
class DenseNet(d2l.Classifier):
    def b1(self):
        return nn.Sequential(
            nn.LazyConv2d(64, kernel_size=7, stride=2, padding=3),
            nn.LazyBatchNorm2d(), nn.ReLU(),
            nn.MaxPool2d(kernel_size=3, stride=2, padding=1))
```

Then, similar to the four modules made up of residual blocks that ResNet uses, DenseNet uses four dense blocks. As with ResNet, we can set the number of convolutional layers used in each dense block. Here, we set it to 4, consistent with the ResNet-18 model in Section 8.6. Furthermore, we set the number of channels (i.e., growth rate) for the convolutional layers in the dense block to 32, so 128 channels will be added to each dense block.

In ResNet, the height and width are reduced between each module by a residual block with a stride of 2. Here, we use the transition layer to halve the height and width and halve the number of channels. Similar to ResNet, a global pooling layer and a fully connected layer are connected at the end to produce the output.

```
@d2l.add_to_class(DenseNet)
def __init__(self, num_channels=64, growth_rate=32, arch=(4, 4, 4, 4),
             lr=0.1, num_classes=10):
    super(DenseNet, self).__init__()
    self.save_hyperparameters()
    self.net = nn.Sequential(self.b1())
    for i, num_convs in enumerate(arch):
        self.net.add_module(f'dense_blk{i+1}', DenseBlock(num_convs,
                                                          growth_rate))
        # The number of output channels in the previous dense block
        num_channels += num_convs * growth_rate
        # A transition layer that halves the number of channels is added
        # between the dense blocks
        if i != len(arch) - 1:
            num_channels //= 2
            self.net.add_module(f'tran_blk{i+1}', transition_block(
                num_channels))
    self.net.add_module('last', nn.Sequential(
        nn.LazyBatchNorm2d(), nn.ReLU(),
        nn.AdaptiveAvgPool2d((1, 1)), nn.Flatten(),
        nn.LazyLinear(num_classes)))
    self.net.apply(d2l.init_cnn)
```

### 8.7.5 Training

Since we are using a deeper network here, in this section, we will reduce the input height and width from 224 to 96 to simplify the computation.

```
model = DenseNet(lr=0.01)
trainer = d2l.Trainer(max_epochs=10, num_gpus=1)
data = d2l.FashionMNIST(batch_size=128, resize=(96, 96))
trainer.fit(model, data)
```

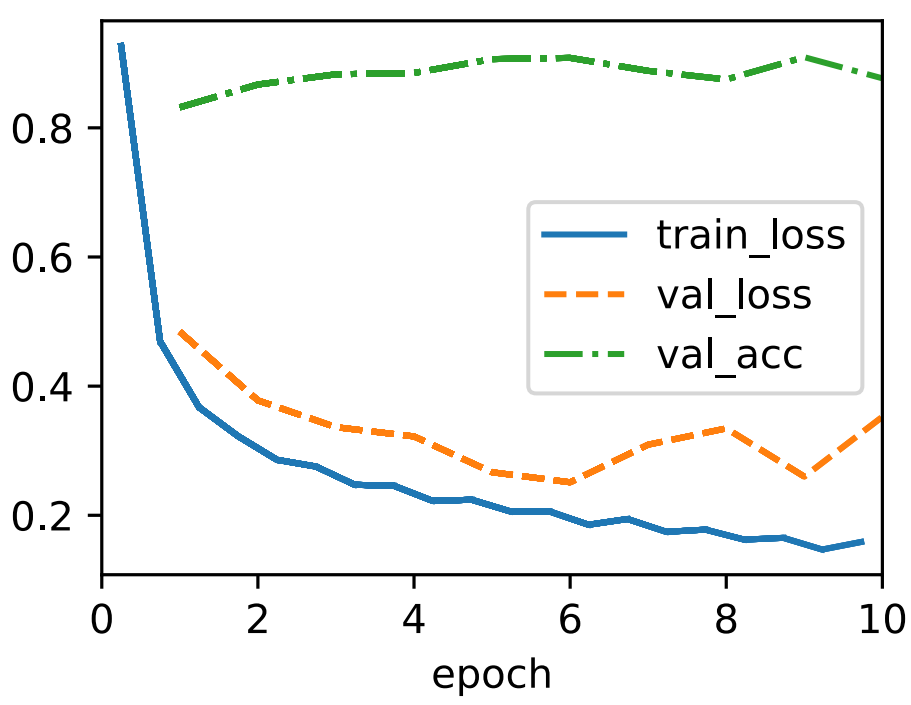

### 8.7.6 Summary and Discussion

The main components that comprise DenseNet are dense blocks and transition layers. For the latter, we need to keep the dimensionality under control when composing the network by adding transition layers that shrink the number of channels again. In terms of cross-layer connections, in contrast to ResNet, where inputs and outputs are added together, DenseNet concatenates inputs and outputs on the channel dimension. Although these concatenation operations reuse features to achieve computational efficiency, unfortunately they lead to heavy GPU memory consumption. As a result, applying DenseNet may require more memory-efficient implementations that may increase training time (Pleiss *et al.*, 2017).

### 8.7.7 Exercises

1. Why do we use average pooling rather than max-pooling in the transition layer?
2. One of the advantages mentioned in the DenseNet paper is that its model parameters are smaller than those of ResNet. Why is this the case?
3. One problem for which DenseNet has been criticized is its high memory consumption.
    1. Is this really the case? Try to change the input shape to $224 \times 224$ to compare the actual GPU memory consumption empirically.

2. Can you think of an alternative means of reducing the memory consumption? How would you need to change the framework?

4. Implement the various DenseNet versions presented in Table 1 of the DenseNet paper (Huang *et al.*, 2017).

5. Design an MLP-based model by applying the DenseNet idea. Apply it to the housing price prediction task in Section 5.7.

Discussions[134].

## 8.8 Designing Convolution Network Architectures

The previous sections have taken us on a tour of modern network design for computer vision. Common to all the work we covered was that it greatly relied on the intuition of scientists. Many of the architectures are heavily informed by human creativity and to a much lesser extent by systematic exploration of the design space that deep networks offer. Nonetheless, this *network engineering* approach has been tremendously successful.

Ever since AlexNet (Section 8.1) beat conventional computer vision models on ImageNet, it has become popular to construct very deep networks by stacking blocks of convolutions, all designed according to the same pattern. In particular, $3 \times 3$ convolutions were popularized by VGG networks (Section 8.2). NiN (Section 8.3) showed that even $1 \times 1$ convolutions could be beneficial by adding local nonlinearities. Moreover, NiN solved the problem of aggregating information at the head of a network by aggregating across all locations. GoogLeNet (Section 8.4) added multiple branches of different convolution width, combining the advantages of VGG and NiN in its Inception block. ResNets (Section 8.6) changed the inductive bias towards the identity mapping (from $f(x) = 0$). This allowed for very deep networks. Almost a decade later, the ResNet design is still popular, a testament to its design. Lastly, ResNeXt (Section 8.6.5) added grouped convolutions, offering a better trade-off between parameters and computation. A precursor to Transformers for vision, the Squeeze-and-Excitation Networks (SENets) allow for efficient information transfer between locations (Hu *et al.*, 2018). This was accomplished by computing a per-channel global attention function.

Up to now we have omitted networks obtained via *neural architecture search* (NAS) (Liu *et al.*, 2018, Zoph and Le, 2016). We chose to do so since their cost is usually enormous, relying on brute-force search, genetic algorithms, reinforcement learning, or some other form of hyperparameter optimization. Given a fixed search space, NAS uses a search strategy to automatically select an architecture based on the returned performance estimation. The outcome of NAS is a single network instance. EfficientNets are a notable outcome of this search (Tan and Le, 2019).

In the following we discuss an idea that is quite different to the quest for the *single best network*. It is computationally relatively inexpensive, it leads to scientific insights on the way, and it is quite effective in terms of the quality of outcomes. Let's review the strategy by Radosavovic *et al.* (2020) to *design network design spaces*. The strategy combines the strength of manual design and NAS. It accomplishes this by operating on *distributions of networks* and optimizing the distributions in a way to obtain good performance for entire families of networks. The outcome of it are *RegNets*, specifically RegNetX and RegNetY, plus a range of guiding principles for the design of performant CNNs.

```
import torch
from torch import nn
from torch.nn import functional as F
from d2l import torch as d2l
```

## 8.8.1 The AnyNet Design Space

The description below closely follows the reasoning in Radosavovic *et al.* (2020) with some abbreviations to make it fit in the scope of the book. To begin, we need a template for the family of networks to explore. One of the commonalities of the designs in this chapter is that the networks consist of a *stem*, a *body* and a *head*. The stem performs initial image processing, often through convolutions with a larger window size. The body consists of multiple blocks, carrying out the bulk of the transformations needed to go from raw images to object representations. Lastly, the head converts this into the desired outputs, such as via a softmax regressor for multiclass classification. The body, in turn, consists of multiple stages, operating on the image at decreasing resolutions. In fact, both the stem and each subsequent stage quarter the spatial resolution. Lastly, each stage consists of one or more blocks. This pattern is common to all networks, from VGG to ResNeXt. Indeed, for the design of generic AnyNet networks, Radosavovic *et al.* (2020) used the ResNeXt block of Fig. 8.6.5.

Let's review the structure outlined in Fig. 8.8.1 in detail. As mentioned, an AnyNet consists of a stem, body, and head. The stem takes as its input RGB images (3 channels), using a $3 \times 3$ convolution with a stride of 2, followed by a batch norm, to halve the resolution from $r \times r$ to $r/2 \times r/2$. Moreover, it generates $c_0$ channels that serve as input to the body.

Since the network is designed to work well with ImageNet images of shape $224 \times 224 \times 3$, the body serves to reduce this to $7 \times 7 \times c_4$ through 4 stages (recall that $224/2^{1+4} = 7$), each with an eventual stride of 2. Lastly, the head employs an entirely standard design via global average pooling, similar to NiN (Section 8.3), followed by a fully connected layer to emit an $n$-dimensional vector for $n$-class classification.

Most of the relevant design decisions are inherent to the body of the network. It proceeds in stages, where each stage is composed of the same type of ResNeXt blocks as we discussed in Section 8.6.5. The design there is again entirely generic: we begin with a block that halves the resolution by using a stride of 2 (the rightmost in Fig. 8.8.1). To match this, the residual branch of the ResNeXt block needs to pass through a $1 \times 1$ convolution. This block is followed by a variable number of additional ResNeXt blocks that leave both resolution and the number of channels unchanged. Note that a common design practice is to add a slight bottleneck in the design of convolutional blocks. As such, with bottleneck ratio $k_i \geq 1$ we afford some number of channels, $c_i/k_i$, within each block for stage $i$ (as the experiments show, this is not really effective and should be skipped). Lastly, since we are dealing with ResNeXt blocks, we also need to pick the number of groups $g_i$ for grouped convolutions at stage $i$.

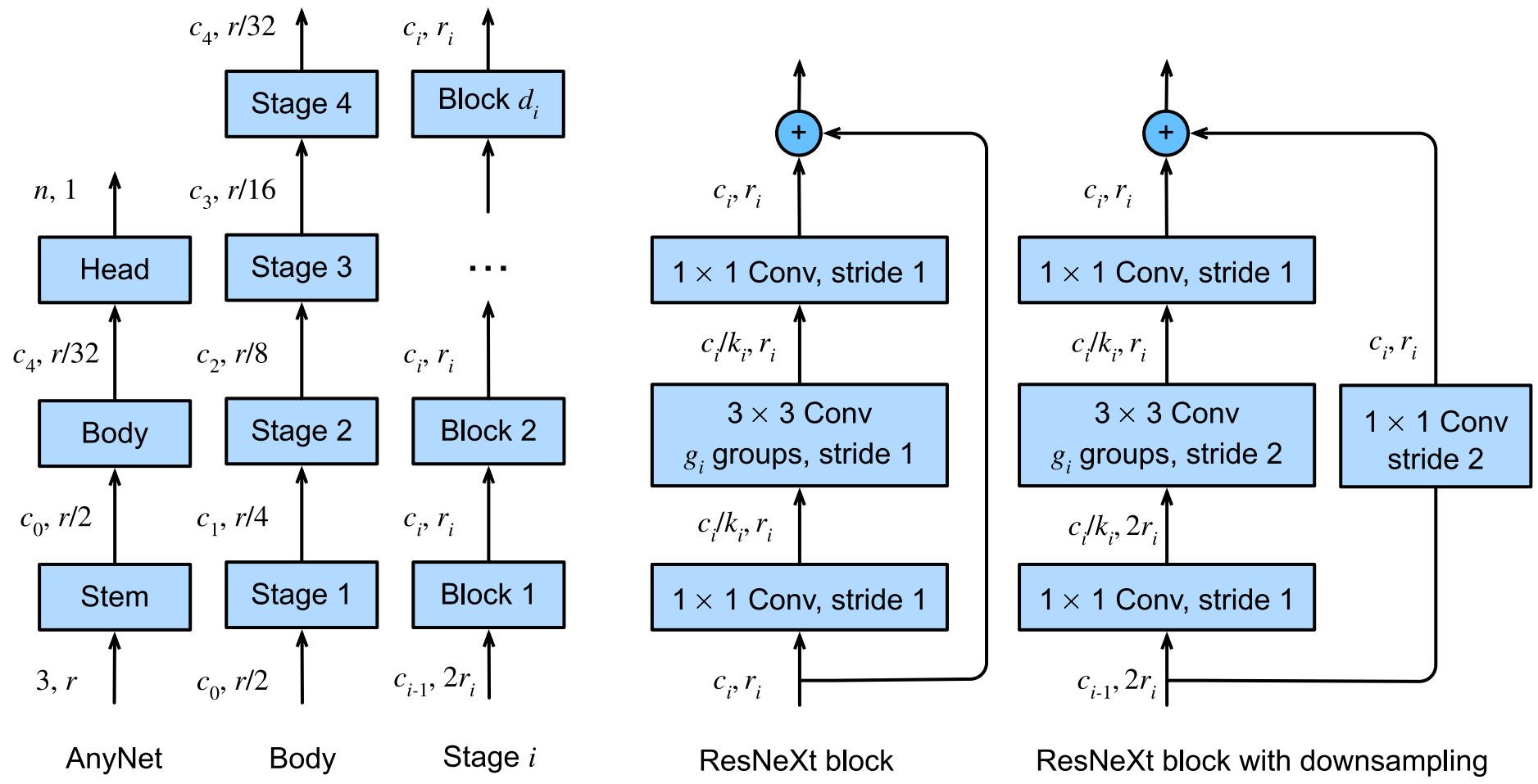

Fig. 8.8.1 The AnyNet design space. The numbers $(c, r)$ along each arrow indicate the number of channels c and the resolution $r \times r$ of the images at that point. From left to right: generic network structure composed of stem, body, and head; body composed of four stages; detailed structure of a stage; two alternative structures for blocks, one without downsampling and one that halves the resolution in each dimension. Design choices include depth $d_i$, the number of output channels $c_i$, the number of groups $g_i$, and bottleneck ratio $k_i$ for any stage $i$.

This seemingly generic design space provides us nonetheless with many parameters: we can set the block width (number of channels) $c_0, \ldots c_4$, the depth (number of blocks) per stage $d_1, \ldots d_4$, the bottleneck ratios $k_1, \ldots k_4$, and the group widths (numbers of groups) $g_1, \ldots g_4$. In total this adds up to 17 parameters, resulting in an unreasonably large number of configurations that would warrant exploring. We need some tools to reduce this huge design space effectively. This is where the conceptual beauty of design spaces comes in. Before we do so, let's implement the generic design first.

```
class AnyNet(d2l.Classifier):
    def stem(self, num_channels):
        return nn.Sequential(
            nn.LazyConv2d(num_channels, kernel_size=3, stride=2, padding=1),
            nn.LazyBatchNorm2d(), nn.ReLU())
```

Each stage consists of `depth` ResNeXt blocks, where `num_channels` specifies the block width. Note that the first block halves the height and width of input images.

```
@d2l.add_to_class(AnyNet)
def stage(self, depth, num_channels, groups, bot_mul):
    blk = []
    for i in range(depth):
        if i == 0:
            blk.append(d2l.ResNeXtBlock(num_channels, groups, bot_mul,
                use_1x1conv=True, strides=2))
        else:
            blk.append(d2l.ResNeXtBlock(num_channels, groups, bot_mul))
    return nn.Sequential(*blk)
```

Putting the network stem, body, and head together, we complete the implementation of AnyNet.

```
@d2l.add_to_class(AnyNet)
def __init__(self, arch, stem_channels, lr=0.1, num_classes=10):
    super(AnyNet, self).__init__()
    self.save_hyperparameters()
    self.net = nn.Sequential(self.stem(stem_channels))
    for i, s in enumerate(arch):
        self.net.add_module(f'stage{i+1}', self.stage(*s))
    self.net.add_module('head', nn.Sequential(
        nn.AdaptiveAvgPool2d((1, 1)), nn.Flatten(),
        nn.LazyLinear(num_classes)))
    self.net.apply(d2l.init_cnn)
```

## 8.8.2 Distributions and Parameters of Design Spaces

As just discussed in Section 8.8.1, parameters of a design space are hyperparameters of networks in that design space. Consider the problem of identifying good parameters in the AnyNet design space. We could try finding the *single best* parameter choice for a given amount of computation (e.g., FLOPs and compute time). If we allowed for even only *two* possible choices for each parameter, we would have to explore $2^{17} = 131072$ combinations to find the best solution. This is clearly infeasible because of its exorbitant cost. Even worse, we do not really learn anything from this exercise in terms of how one should design a network. Next time we add, say, an X-stage, or a shift operation, or similar, we would need to start from scratch. Even worse, due to the stochasticity in training (rounding, shuffling, bit errors), no two runs are likely to produce exactly the same results. A better strategy would be to try to determine general guidelines of how the choices of parameters should be related. For instance, the bottleneck ratio, the number of channels, blocks, groups, or their change between layers should ideally be governed by a collection of simple rules. The approach in Radosavovic *et al.* (2019) relies on the following four assumptions:

1. We assume that general design principles actually exist, so that many networks satisfying these requirements should offer good performance. Consequently, identifying a *distribution* over networks can be a sensible strategy. In other words, we assume that there are many good needles in the haystack.

2. We need not train networks to convergence before we can assess whether a network is good. Instead, it is sufficient to use the intermediate results as reliable guidance for final accuracy. Using (approximate) proxies to optimize an objective is referred to as multi-fidelity optimization (Forrester *et al.*, 2007). Consequently, design optimization is carried out, based on the accuracy achieved after only a few passes through the dataset, reducing the cost significantly.

3. Results obtained at a smaller scale (for smaller networks) generalize to larger ones. Consequently, optimization is carried out for networks that are structurally similar, but with a smaller number of blocks, fewer channels, etc. Only in the end will we need to verify that the so-found networks also offer good performance at scale.

4. Aspects of the design can be approximately factorized so that it is possible to infer their effect on the quality of the outcome somewhat independently. In other words, the optimization problem is moderately easy.

These assumptions allow us to test many networks cheaply. In particular, we can *sample* uniformly from the space of configurations and evaluate their performance. Subsequently, we can evaluate the quality of the choice of parameters by reviewing the *distribution* of error/accuracy that can be achieved with said networks. Denote by $F(e)$ the cumulative distribution function (CDF) for errors committed by networks of a given design space, drawn using probability disribution $p$. That is,

$$F(e, p) \stackrel{\text{def}}{=} P_{\text{net} \sim p} \{e(\text{net}) \leq e\}. \tag{8.8.1}$$

Our goal is now to find a distribution $p$ over *networks* such that most networks have a very low error rate and where the support of $p$ is concise. Of course, this is computationally infeasible to perform accurately. We resort to a sample of networks $\mathcal{Z} \stackrel{\text{def}}{=} \{\text{net}_1, \ldots \text{net}_n\}$ (with errors $e_1, \ldots, e_n$, respectively) from $p$ and use the empirical CDF $\hat{F}(e, \mathcal{Z})$ instead:

$$\hat{F}(e, \mathcal{Z}) = \frac{1}{n} \sum_{i=1}^{n} \mathbf{1}(e_i \leq e). \tag{8.8.2}$$

Whenever the CDF for one set of choices majorizes (or matches) another CDF it follows that its choice of parameters is superior (or indifferent). Accordingly Radosavovic *et al.* (2020) experimented with a shared network bottleneck ratio $k_i = k$ for all stages $i$ of the network. This gets rid of three of the four parameters governing the bottleneck ratio. To assess whether this (negatively) affects the performance one can draw networks from the constrained and from the unconstrained distribution and compare the corresonding CDFs. It turns out that this constraint does not affect the accuracy of the distribution of networks at all, as can be seen in the first panel of Fig. 8.8.2. Likewise, we could choose to pick the same group width $g_i = g$ occurring at the various stages of the network. Again, this does not affect performance, as can be seen in the second panel of Fig. 8.8.2. Both steps combined reduce the number of free parameters by six.

Next we look for ways to reduce the multitude of potential choices for width and depth of the stages. It is a reasonable assumption that, as we go deeper, the number of channels should increase, i.e., $c_i \geq c_{i-1}$ ($w_{i+1} \geq w_i$ per their notation in Fig. 8.8.2), yielding AnyNetX$_D$.

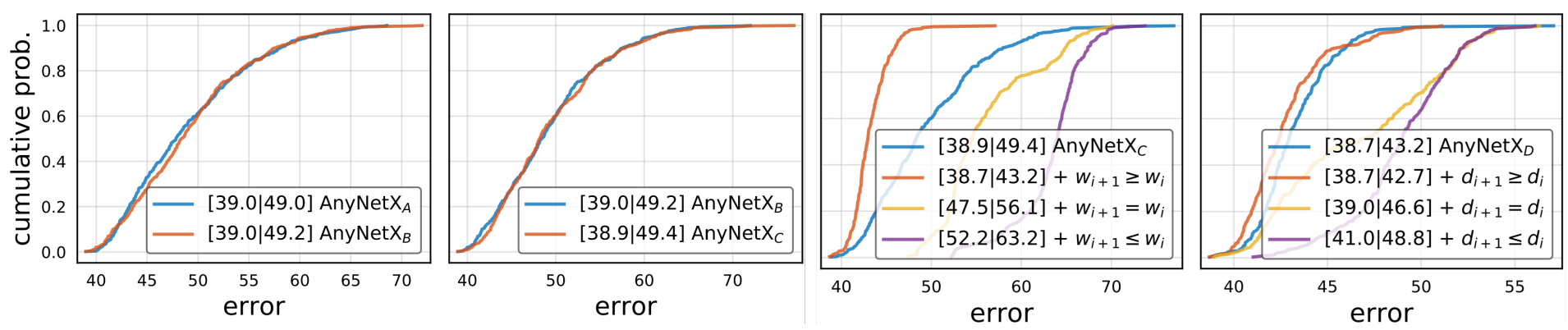

Fig. 8.8.2 Comparing error empirical distribution functions of design spaces. AnyNet$_A$ is the original design space; AnyNet$_B$ ties the bottleneck ratios, AnyNet$_C$ also ties group widths, AnyNet$_D$ increases the network depth across stages. From left to right: (i) tying bottleneck ratios has no effect on performance; (ii) tying group widths has no effect on performance; (iii) increasing network widths (channels) across stages improves performance; (iv) increasing network depths across stages improves performance. Figure courtesy of Radosavovic et al. (2020).

Likewise, it is equally reasonable to assume that as the stages progress, they should become deeper, i.e., $d_i \geq d_{i-1}$, yielding AnyNetX$_E$. This can be experimentally verified in the third and fourth panel of Fig. 8.8.2, respectively.

## 8.8.3 RegNet

The resulting AnyNetX$_E$ design space consists of simple networks following easy-to-interpret design principles:

- Share the bottleneck ratio $k_i = k$ for all stages $i$;
- Share the group width $g_i = g$ for all stages $i$;
- Increase network width across stages: $c_i \leq c_{i+1}$;
- Increase network depth across stages: $d_i \leq d_{i+1}$.

This leaves us with a final set of choices: how to pick the specific values for the above parameters of the eventual AnyNetX$_E$ design space. By studying the best-performing networks from the distribution in AnyNetX$_E$ one can observe the following: the width of the network ideally increases linearly with the block index across the network, i.e., $c_j \approx c_0 + c_a j$, where $j$ is the block index and slope $c_a > 0$. Given that we get to choose a different block width only per stage, we arrive at a piecewise constant function, engineered to match this dependence. Furthermore, experiments also show that a bottleneck ratio of $k = 1$ performs best, i.e., we are advised not to use bottlenecks at all.

We recommend the interested reader reviews further details in the design of specific networks for different amounts of computation by perusing Radosavovic *et al.* (2020). For instance, an effective 32-layer RegNetX variant is given by $k = 1$ (no bottleneck), $g = 16$ (group width is 16), $c_1 = 32$ and $c_2 = 80$ channels for the first and second stage, respectively, chosen to be $d_1 = 4$ and $d_2 = 6$ blocks deep. The astonishing insight from the design is that it still applies, even when investigating networks at a larger scale. Even better, it even holds for Squeeze-and-Excitation (SE) network designs (RegNetY) that have a global channel activation (Hu *et al.*, 2018).

```
class RegNetX32(AnyNet):
    def __init__(self, lr=0.1, num_classes=10):
        stem_channels, groups, bot_mul = 32, 16, 1
        depths, channels = (4, 6), (32, 80)
        super().__init__(
            ((depths[0], channels[0], groups, bot_mul),
             (depths[1], channels[1], groups, bot_mul)),
            stem_channels, lr, num_classes)
```

We can see that each RegNetX stage progressively reduces resolution and increases output channels.

```
RegNetX32().layer_summary((1, 1, 96, 96))
```

```
Sequential output shape:     torch.Size([1, 32, 48, 48])
Sequential output shape:     torch.Size([1, 32, 24, 24])
Sequential output shape:     torch.Size([1, 80, 12, 12])
Sequential output shape:     torch.Size([1, 10])
```

### 8.8.4 Training

Training the 32-layer RegNetX on the Fashion-MNIST dataset is just like before.

```
model = RegNetX32(lr=0.05)
trainer = d2l.Trainer(max_epochs=10, num_gpus=1)
data = d2l.FashionMNIST(batch_size=128, resize=(96, 96))
trainer.fit(model, data)
```

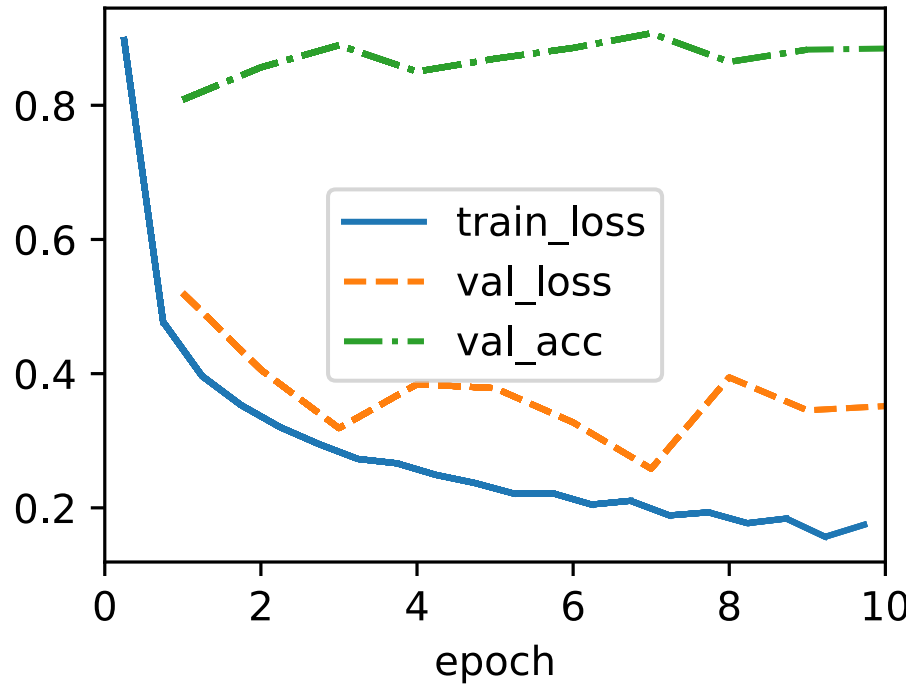

### 8.8.5 Discussion

With desirable inductive biases (assumptions or preferences) like locality and translation invariance (Section 7.1) for vision, CNNs have been the dominant architectures in this area. This remained the case from LeNet up until Transformers (Section 11.7) (Dosovitskiy *et al.*, 2021, Touvron *et al.*, 2021) started surpassing CNNs in terms of accuracy. While much of the recent progress in terms of vision Transformers *can* be backported into CNNs (Liu *et al.*, 2022), it is only possible at a higher computational cost. Just as importantly, recent hardware optimizations (NVIDIA Ampere and Hopper) have only widened the gap in favor of Transformers.

It is worth noting that Transformers have a significantly lower degree of inductive bias towards locality and translation invariance than CNNs. That learned structures prevailed is due, not least, to the availability of large image collections, such as LAION-400m and LAION-5B (Schuhmann *et al.*, 2022) with up to 5 billion images. Quite surprisingly, some of the more relevant work in this context even includes MLPs (Tolstikhin *et al.*, 2021).

In sum, vision Transformers (Section 11.8) by now lead in terms of state-of-the-art performance in large-scale image classification, showing that *scalability trumps inductive biases* (Dosovitskiy *et al.*, 2021). This includes pretraining large-scale Transformers (Section 11.9) with multi-head self-attention (Section 11.5). We invite the readers to dive into these chapters for a much more detailed discussion.

### 8.8.6 Exercises

1. Increase the number of stages to four. Can you design a deeper RegNetX that performs better?
2. De-ResNeXt-ify RegNets by replacing the ResNeXt block with the ResNet block. How does your new model perform?
3. Implement multiple instances of a "VioNet" family by *violating* the design principles of RegNetX. How do they perform? Which of $(d_i, c_i, g_i, b_i)$ is the most important factor?
4. Your goal is to design the "perfect" MLP. Can you use the design principles introduced above to find good architectures? Is it possible to extrapolate from small to large networks?

Discussions[135].

# 9 Recurrent Neural Networks

Up until now, we have focused primarily on fixed-length data. When introducing linear and logistic regression in Chapter 3 and Chapter 4 and multilayer perceptrons in Chapter 5, we were happy to assume that each feature vector $\mathbf{x}_i$ consisted of a fixed number of components $x_1, \ldots, x_d$, where each numerical feature $x_j$ corresponded to a particular attribute. These datasets are sometimes called *tabular*, because they can be arranged in tables, where each example $i$ gets its own row, and each attribute gets its own column. Crucially, with tabular data, we seldom assume any particular structure over the columns.

Subsequently, in Chapter 7, we moved on to image data, where inputs consist of the raw pixel values at each coordinate in an image. Image data hardly fitted the bill of a protypical tabular dataset. There, we needed to call upon convolutional neural networks (CNNs) to handle the hierarchical structure and invariances. However, our data were still of fixed length. Every Fashion-MNIST image is represented as a $28 \times 28$ grid of pixel values. Moreover, our goal was to develop a model that looked at just one image and then outputted a single prediction. But what should we do when faced with a sequence of images, as in a video, or when tasked with producing a sequentially structured prediction, as in the case of image captioning?

A great many learning tasks require dealing with sequential data. Image captioning, speech synthesis, and music generation all require that models produce outputs consisting of sequences. In other domains, such as time series prediction, video analysis, and musical information retrieval, a model must learn from inputs that are sequences. These demands often arise simultaneously: tasks such as translating passages of text from one natural language to another, engaging in dialogue, or controlling a robot, demand that models both ingest and output sequentially structured data.

Recurrent neural networks (RNNs) are deep learning models that capture the dynamics of sequences via *recurrent* connections, which can be thought of as cycles in the network of nodes. This might seem counterintuitive at first. After all, it is the feedforward nature of neural networks that makes the order of computation unambiguous. However, recurrent edges are defined in a precise way that ensures that no such ambiguity can arise. Recurrent neural networks are *unrolled* across time steps (or sequence steps), with the *same* underlying parameters applied at each step. While the standard connections are applied *synchronously* to propagate each layer's activations to the subsequent layer *at the same time step*, the recurrent connections are *dynamic*, passing information across adjacent time steps. As the unfolded view in Fig. 9.1 reveals, RNNs can be thought of as feedforward neural networks where each layer's parameters (both conventional and recurrent) are shared across time steps.

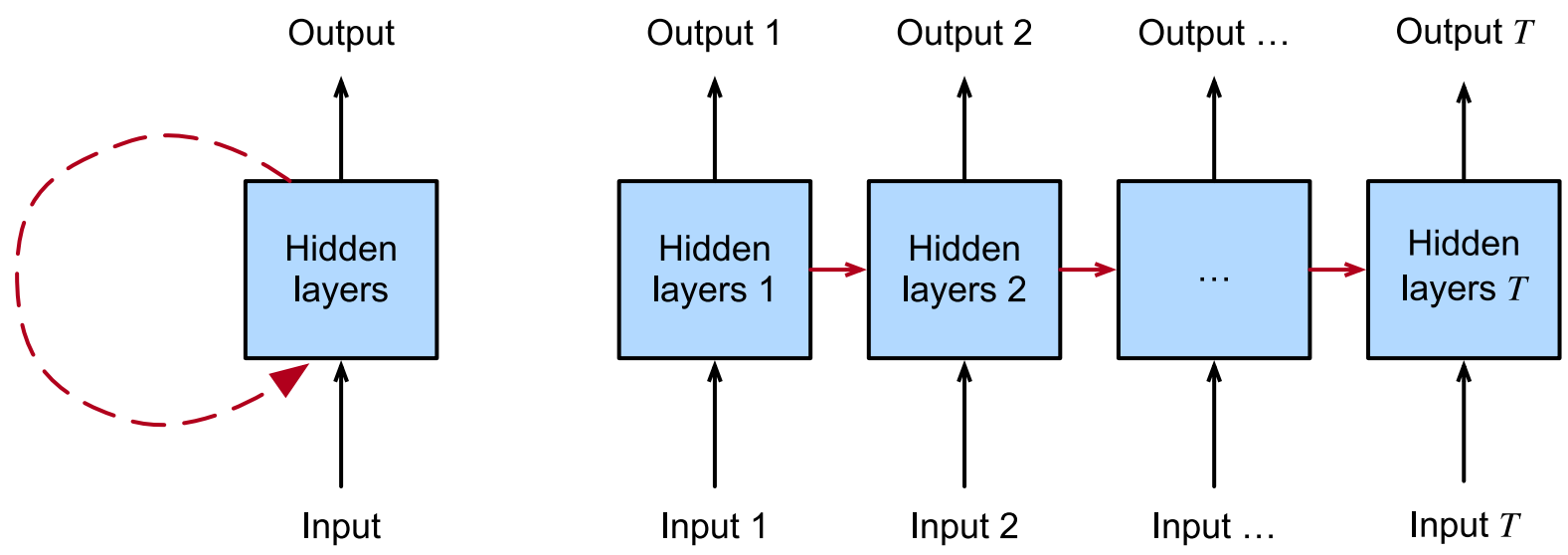

Fig. 9.1 On the left recurrent connections are depicted via cyclic edges. On the right, we unfold the RNN over time steps. Here, recurrent edges span adjacent time steps, while conventional connections are computed synchronously.

Like neural networks more broadly, RNNs have a long discipline-spanning history, originating as models of the brain popularized by cognitive scientists and subsequently adopted as practical modeling tools employed by the machine learning community. As we do for deep learning more broadly, in this book we adopt the machine learning perspective, focusing on RNNs as practical tools that rose to popularity in the 2010s owing to breakthrough results on such diverse tasks as handwriting recognition (Graves *et al.*, 2008), machine translation (Sutskever *et al.*, 2014), and recognizing medical diagnoses (Lipton *et al.*, 2016). We point the reader interested in more background material to a publicly available comprehensive review (Lipton *et al.*, 2015). We also note that sequentiality is not unique to RNNs. For example, the CNNs that we already introduced can be adapted to handle data of varying length, e.g., images of varying resolution. Moreover, RNNs have recently ceded considerable market share to Transformer models, which will be covered in Chapter 11. However, RNNs rose to prominence as the default models for handling complex sequential structure in deep learning, and remain staple models for sequential modeling to this day. The stories of RNNs and of sequence modeling are inextricably linked, and this is as much a chapter about the ABCs of sequence modeling problems as it is a chapter about RNNs.

One key insight paved the way for a revolution in sequence modeling. While the inputs and targets for many fundamental tasks in machine learning cannot easily be represented as fixed-length vectors, they can often nevertheless be represented as varying-length sequences of fixed-length vectors. For example, documents can be represented as sequences of words; medical records can often be represented as sequences of events (encounters, medications, procedures, lab tests, diagnoses); videos can be represented as varying-length sequences of still images.

While sequence models have popped up in numerous application areas, basic research in the area has been driven predominantly by advances on core tasks in natural language processing. Thus, throughout this chapter, we will focus our exposition and examples on text data. If you get the hang of these examples, then applying the models to other data modalities should be relatively straightforward. In the next few sections, we introduce basic notation for sequences and some evaluation measures for assessing the quality of sequentially structured model outputs. After that, we discuss basic concepts of a language model and use this dis-cussion to motivate our first RNN models. Finally, we describe the method for calculating gradients when backpropagating through RNNs and explore some challenges that are often encountered when training such networks, motivating the modern RNN architectures that will follow in Chapter 10.

## 9.1 Working with Sequences

Up until now, we have focused on models whose inputs consisted of a single feature vector $\mathbf{x} \in \mathbb{R}^d$. The main change of perspective when developing models capable of processing sequences is that we now focus on inputs that consist of an ordered list of feature vectors $\mathbf{x}_1, \ldots, \mathbf{x}_T$, where each feature vector $\mathbf{x}_t$ is indexed by a time step $t \in \mathbb{Z}^+$ lying in $\mathbb{R}^d$.

Some datasets consist of a single massive sequence. Consider, for example, the extremely long streams of sensor readings that might be available to climate scientists. In such cases, we might create training datasets by randomly sampling subsequences of some predetermined length. More often, our data arrives as a collection of sequences. Consider the following examples: (i) a collection of documents, each represented as its own sequence of words, and each having its own length $T_i$; (ii) sequence representation of patient stays in the hospital, where each stay consists of a number of events and the sequence length depends roughly on the length of the stay.

Previously, when dealing with individual inputs, we assumed that they were sampled independently from the same underlying distribution $P(X)$. While we still assume that entire sequences (e.g., entire documents or patient trajectories) are sampled independently, we cannot assume that the data arriving at each time step are independent of each other. For example, the words that likely to appear later in a document depend heavily on words occurring earlier in the document. The medicine a patient is likely to receive on the 10th day of a hospital visit depends heavily on what transpired in the previous nine days.

This should come as no surprise. If we did not believe that the elements in a sequence were related, we would not have bothered to model them as a sequence in the first place. Consider the usefulness of the auto-fill features that are popular on search tools and modern email clients. They are useful precisely because it is often possible to predict (imperfectly, but better than random guessing) what the likely continuations of a sequence might be, given some initial prefix. For most sequence models, we do not require independence, or even stationarity, of our sequences. Instead, we require only that the sequences themselves are sampled from some fixed underlying distribution over entire sequences.

This flexible approach allows for such phenomena as (i) documents looking significantly different at the beginning than at the end; or (ii) patient status evolving either towards recovery or towards death over the course of a hospital stay; or (iii) customer taste evolving in predictable ways over the course of continued interaction with a recommender system.

We sometimes wish to predict a fixed target $y$ given sequentially structured input (e.g., sentiment classification based on a movie review). At other times, we wish to predict a sequentially structured target $(y_1, \ldots, y_T)$ given a fixed input (e.g., image captioning). Still other times, our goal is to predict sequentially structured targets based on sequentially structured inputs (e.g., machine translation or video captioning). Such sequence-to-sequence tasks take two forms: (i) *aligned*: where the input at each time step aligns with a corresponding target (e.g., part of speech tagging); (ii) *unaligned*: where the input and target do not necessarily exhibit a step-for-step correspondence (e.g., machine translation).

Before we worry about handling targets of any kind, we can tackle the most straightforward problem: unsupervised density modeling (also called *sequence modeling*). Here, given a collection of sequences, our goal is to estimate the probability mass function that tells us how likely we are to see any given sequence, i.e., $p(\mathbf{x}_1, \ldots, \mathbf{x}_T)$.

```
%matplotlib inline
import torch
from torch import nn
from d2l import torch as d2l
```

## 9.1.1 Autoregressive Models

Before introducing specialized neural networks designed to handle sequentially structured data, let's take a look at some actual sequence data and build up some basic intuitions and statistical tools. In particular, we will focus on stock price data from the FTSE 100 index (Fig. 9.1.1). At each *time step* $t \in \mathbb{Z}^+$, we observe the price, $x_t$, of the index at that time.

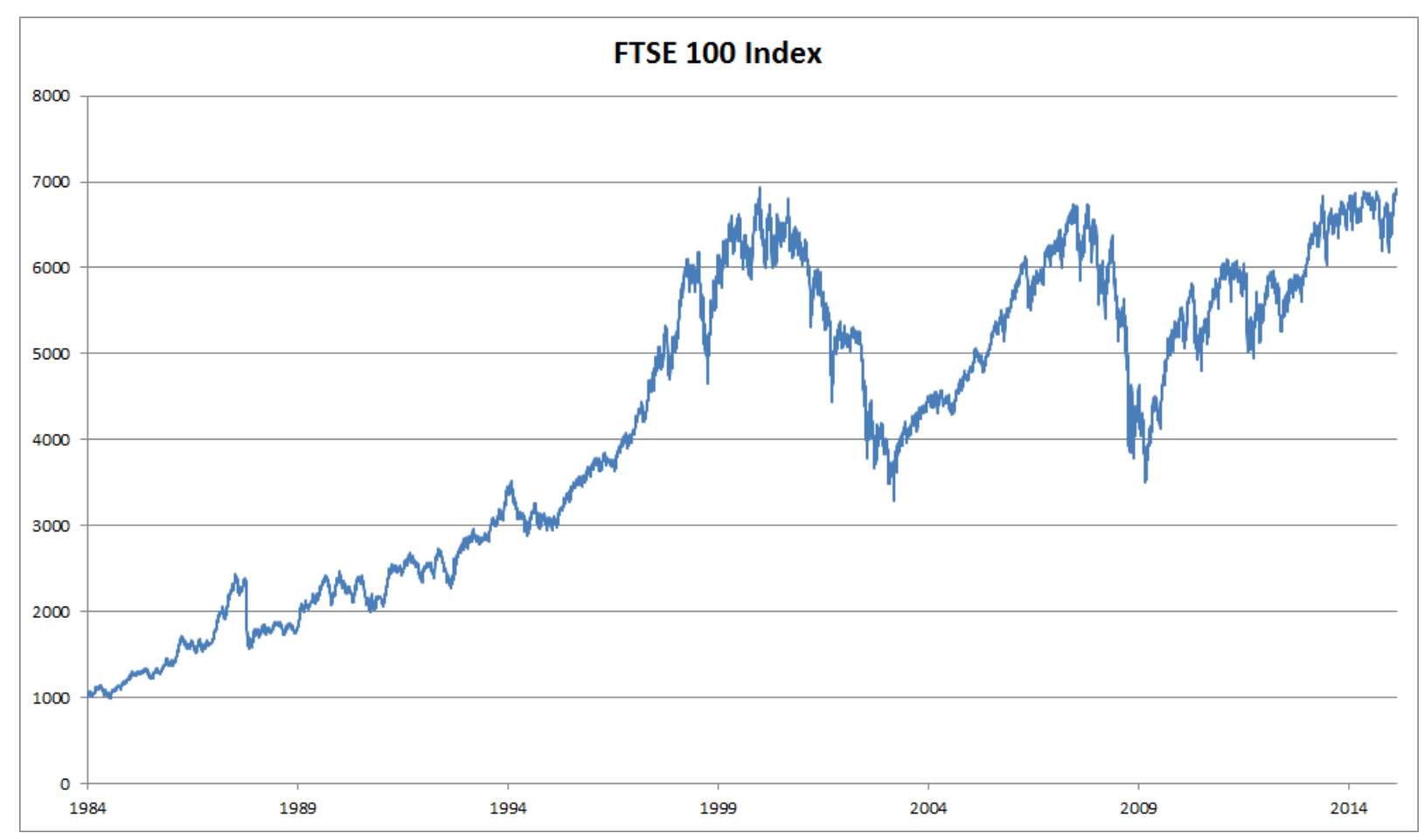

Fig. 9.1.1 FTSE 100 index over about 30 years.

Now suppose that a trader would like to make short-term trades, strategically getting into or out of the index, depending on whether they believe that it will rise or decline in the subsequent time step. Absent any other features (news, financial reporting data, etc.), the only available signal for predicting the subsequent value is the history of prices to date. The trader is thus interested in knowing the probability distribution

$$P(x_t \mid x_{t-1}, \ldots, x_1) \tag{9.1.1}$$

over prices that the index might take in the subsequent time step. While estimating the entire distribution over a continuously valued random variable can be difficult, the trader would be happy to focus on a few key statistics of the distribution, particularly the expected value and the variance. One simple strategy for estimating the conditional expectation

$$\mathbb{E}[(x_t \mid x_{t-1}, \ldots, x_1)], \tag{9.1.2}$$

would be to apply a linear regression model (recall Section 3.1). Such models that regress the value of a signal on the previous values of that same signal are naturally called *autoregressive models*. There is just one major problem: the number of inputs, $x_{t-1}, \ldots, x_1$ varies, depending on $t$. In other words, the number of inputs increases with the amount of data that we encounter. Thus if we want to treat our historical data as a training set, we are left with the problem that each example has a different number of features. Much of what follows in this chapter will revolve around techniques for overcoming these challenges when engaging in such *autoregressive* modeling problems where the object of interest is $P(x_t \mid x_{t-1}, \ldots, x_1)$ or some statistic(s) of this distribution.

A few strategies recur frequently. First of all, we might believe that although long sequences $x_{t-1}, \ldots, x_1$ are available, it may not be necessary to look back so far in the history when predicting the near future. In this case we might content ourselves to condition on some window of length $\tau$ and only use $x_{t-1}, \ldots, x_{t-\tau}$ observations. The immediate benefit is that now the number of arguments is always the same, at least for $t > \tau$. This allows us to train any linear model or deep network that requires fixed-length vectors as inputs. Second, we might develop models that maintain some summary $h_t$ of the past observations (see Fig. 9.1.2) and at the same time update $h_t$ in addition to the prediction $\hat{x}_t$. This leads to models that estimate not only $x_t$ with $\hat{x}_t = P(x_t \mid h_t)$ but also updates of the form $h_t = g(h_{t-1}, x_{t-1})$. Since $h_t$ is never observed, these models are also called *latent autoregressive models*.

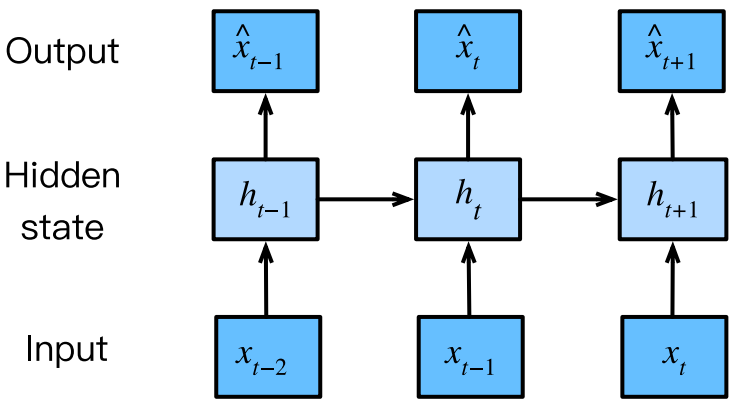

Fig. 9.1.2 A latent autoregressive model.

To construct training data from historical data, one typically creates examples by sampling windows randomly. In general, we do not expect time to stand still. However, we often assume that while the specific values of $x_t$ might change, the dynamics according to which each subsequent observation is generated given the previous observations do not. Statisticians call dynamics that do not change *stationary*.

### 9.1.2 Sequence Models

Sometimes, especially when working with language, we wish to estimate the joint probability of an entire sequence. This is a common task when working with sequences composed of discrete *tokens*, such as words. Generally, these estimated functions are called *sequence models* and for natural language data, they are called *language models*. The field of sequence modeling has been driven so much by natural language processing, that we often describe sequence models as "language models", even when dealing with non-language data. Language models prove useful for all sorts of reasons. Sometimes we want to evaluate the likelihood of sentences. For example, we might wish to compare the naturalness of two candidate outputs generated by a machine translation system or by a speech recognition system. But language modeling gives us not only the capacity to *evaluate* likelihood, but the ability to *sample* sequences, and even to optimize for the most likely sequences.

While language modeling might not, at first glance, look like an autoregressive problem, we can reduce language modeling to autoregressive prediction by decomposing the joint density of a sequence $p(x_1, \ldots, x_T)$ into the product of conditional densities in a left-to-right fashion by applying the chain rule of probability:

$$P(x_1, \ldots, x_T) = P(x_1) \prod_{t=2}^{T} P(x_t \mid x_{t-1}, \ldots, x_1). \tag{9.1.3}$$

Note that if we are working with discrete signals such as words, then the autoregressive model must be a probabilistic classifier, outputting a full probability distribution over the vocabulary for whatever word will come next, given the leftwards context.

#### Markov Models

Now suppose that we wish to employ the strategy mentioned above, where we condition only on the $\tau$ previous time steps, i.e., $x_{t-1}, \ldots, x_{t-\tau}$, rather than the entire sequence history $x_{t-1}, \ldots, x_1$. Whenever we can throw away the history beyond the previous $\tau$ steps without any loss in predictive power, we say that the sequence satisfies a *Markov condition*, i.e., *that the future is conditionally independent of the past, given the recent history*. When $\tau = 1$, we say that the data is characterized by a *first-order Markov model*, and when $\tau = k$, we say that the data is characterized by a $k^{\text{th}}$-order Markov model. For when the first-order Markov condition holds ($\tau = 1$) the factorization of our joint probability becomes a product of probabilities of each word given the previous *word*:

$$P(x_1, \ldots, x_T) = P(x_1) \prod_{t=2}^{T} P(x_t \mid x_{t-1}). \tag{9.1.4}$$

We often find it useful to work with models that proceed as though a Markov condition were satisfied, even when we know that this is only *approximately* true. With real text documents we continue to gain information as we include more and more leftwards context. But these gains diminish rapidly. Thus, sometimes we compromise, obviating computational and statistical difficulties by training models whose validity depends on a $k^{\text{th}}$-order Markov condition. Even today's massive RNN- and Transformer-based language models seldom incorporate more than thousands of words of context.

With discrete data, a true Markov model simply counts the number of times that each word has occurred in each context, producing the relative frequency estimate of $P(x_t \mid x_{t-1})$. Whenever the data assumes only discrete values (as in language), the most likely sequence of words can be computed efficiently using dynamic programming.

### The Order of Decoding

You may be wondering why we represented the factorization of a text sequence $P(x_1, \ldots, x_T)$ as a left-to-right chain of conditional probabilities. Why not right-to-left or some other, seemingly random order? In principle, there is nothing wrong with unfolding $P(x_1, \ldots, x_T)$ in reverse order. The result is a valid factorization:

$$P(x_1, \ldots, x_T) = P(x_T) \prod_{t=T-1}^{1} P(x_t \mid x_{t+1}, \ldots, x_T). \tag{9.1.5}$$

However, there are many reasons why factorizing text in the same direction in which we read it (left-to-right for most languages, but right-to-left for Arabic and Hebrew) is preferred for the task of language modeling. First, this is just a more natural direction for us to think about. After all we all read text every day, and this process is guided by our ability to anticipate which words and phrases are likely to come next. Just think of how many times you have completed someone else's sentence. Thus, even if we had no other reason to prefer such in-order decodings, they would be useful if only because we have better intuitions for what should be likely when predicting in this order.

Second, by factorizing in order, we can assign probabilities to arbitrarily long sequences using the same language model. To convert a probability over steps 1 through $t$ into one that extends to word $t+1$ we simply multiply by the conditional probability of the additional token given the previous ones: $P(x_{t+1}, \ldots, x_1) = P(x_t, \ldots, x_1) \cdot P(x_{t+1} \mid x_t, \ldots, x_1)$.

Third, we have stronger predictive models for predicting adjacent words than words at arbitrary other locations. While all orders of factorization are valid, they do not necessarily all represent equally easy predictive modeling problems. This is true not only for language, but for other kinds of data as well, e.g., when the data is causally structured. For example, we believe that future events cannot influence the past. Hence, if we change $x_t$, we may be able to influence what happens for $x_{t+1}$ going forward but not the converse. That is, if we change $x_t$, the distribution over past events will not change. In some contexts, this makes it easier to predict $P(x_{t+1} \mid x_t)$ than to predict $P(x_t \mid x_{t+1})$. For instance, in some cases, we can find $x_{t+1} = f(x_t) + \epsilon$ for some additive noise $\epsilon$, whereas the converse is not true (Hoyer *et al.*, 2009). This is great news, since it is typically the forward direction that we are interested in estimating. The book by Peters *et al.* (2017) contains more on this topic. We barely scratch the surface of it.

## 9.1.3 Training

Before we focus our attention on text data, let's first try this out with some continuous-valued synthetic data.

Here, our 1000 synthetic data will follow the trigonometric `sin` function, applied to 0.01 times the time step. To make the problem a little more interesting, we corrupt each sample with additive noise. From this sequence we extract training examples, each consisting of features and a label.

```
class Data(d2l.DataModule):
    def __init__(self, batch_size=16, T=1000, num_train=600, tau=4):
        self.save_hyperparameters()
        self.time = torch.arange(1, T + 1, dtype=torch.float32)
        self.x = torch.sin(0.01 * self.time) + torch.randn(T) * 0.2
```

```
data = Data()
d2l.plot(data.time, data.x, 'time', 'x', xlim=[1, 1000], figsize=(6, 3))
```

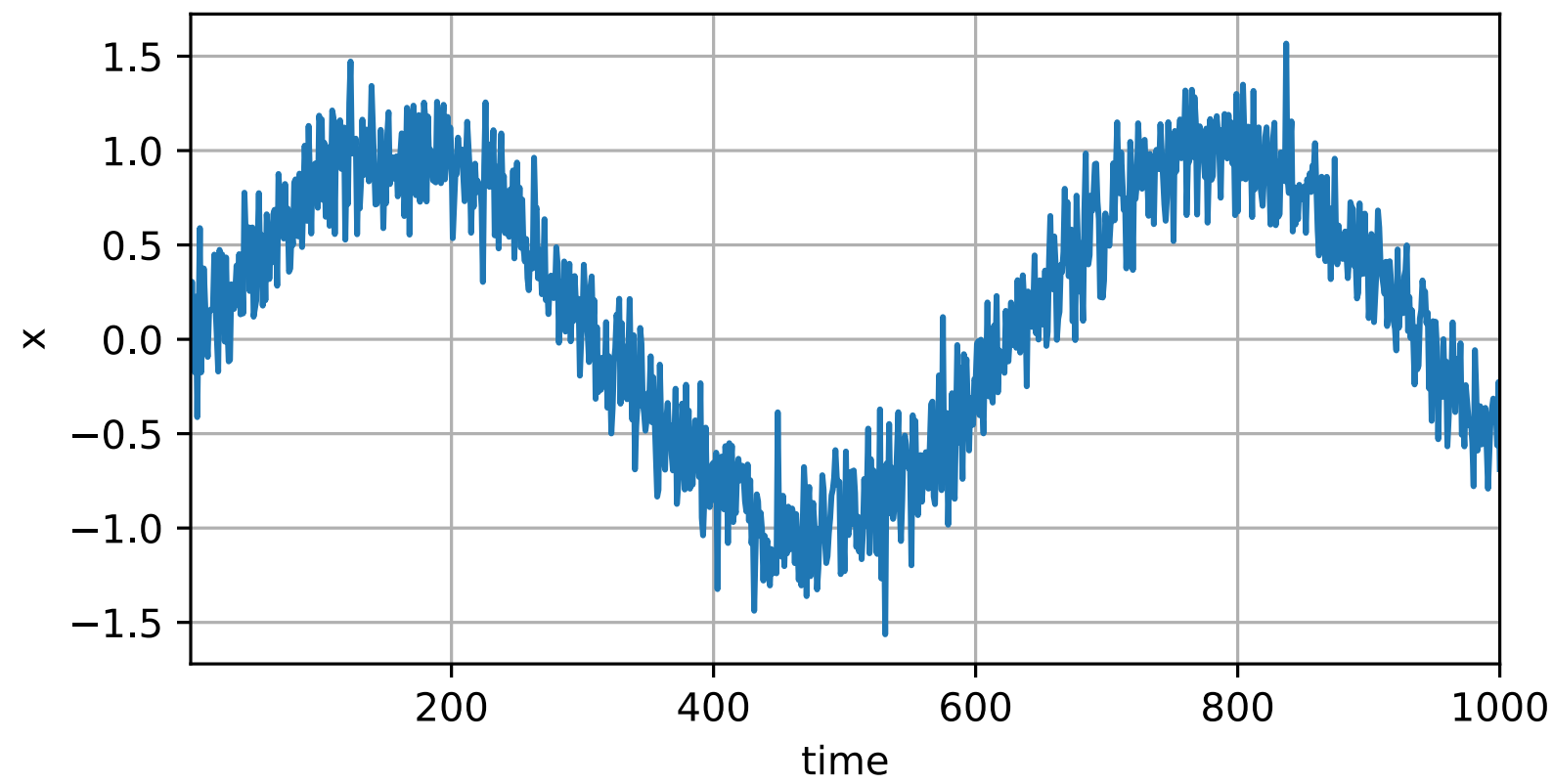

To begin, we try a model that acts as if the data satisfied a $\tau^{\textrm{th}}$-order Markov condition, and thus predicts $x_t$ using only the past $\tau$ observations. Thus for each time step we have an example with label $y = x_t$ and features $\mathbf{x}_t = [x_{t-\tau}, \ldots, x_{t-1}]$. The astute reader might have noticed that this results in $1000-\tau$ examples, since we lack sufficient history for $y_1, \ldots, y_\tau$. While we could pad the first $\tau$ sequences with zeros, to keep things simple, we drop them for now. The resulting dataset contains $T - \tau$ examples, where each input to the model has sequence length $\tau$. We create a data iterator on the first 600 examples, covering a period of the sin function.

```
@d2l.add_to_class(Data)
def get_dataloader(self, train):
    features = [self.x[i : self.T-self.tau+i] for i in range(self.tau)]
    self.features = torch.stack(features, 1)
    self.labels = self.x[self.tau:].reshape((-1, 1))
    i = slice(0, self.num_train) if train else slice(self.num_train, None)
    return self.get_tensorloader([self.features, self.labels], train, i)
```

In this example our model will be a standard linear regression.

```
model = d2l.LinearRegression(lr=0.01)
trainer = d2l.Trainer(max_epochs=5)
trainer.fit(model, data)
```

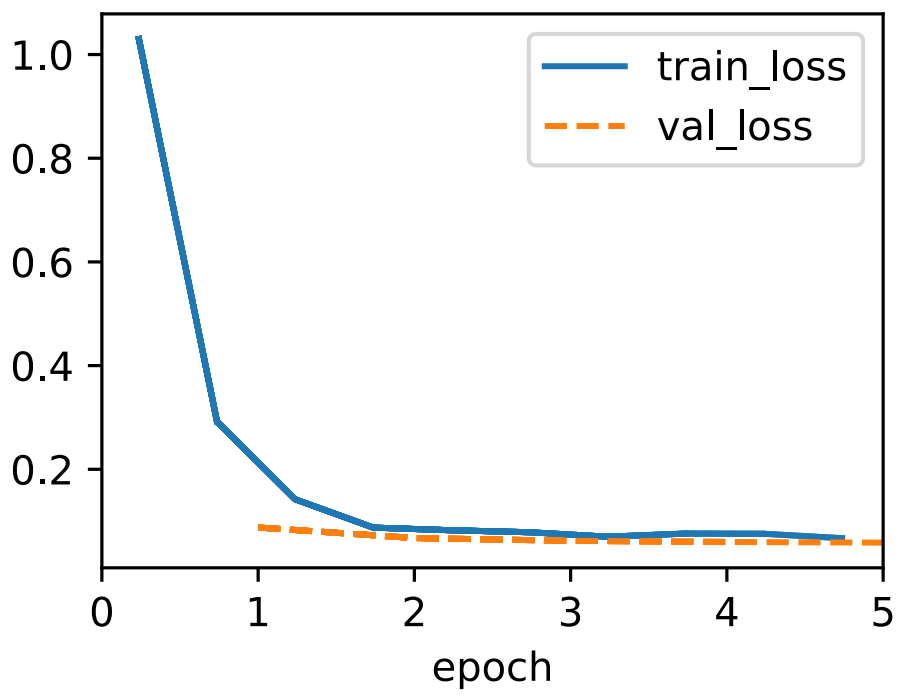

## 9.1.4 Prediction

To evaluate our model, we first check how well it performs at one-step-ahead prediction.

```
onestep_preds = model(data.features).detach().numpy()
d2l.plot(data.time[data.tau:], [data.labels, onestep_preds], 'time', 'x',
         legend=['labels', '1-step preds'], figsize=(6, 3))
```

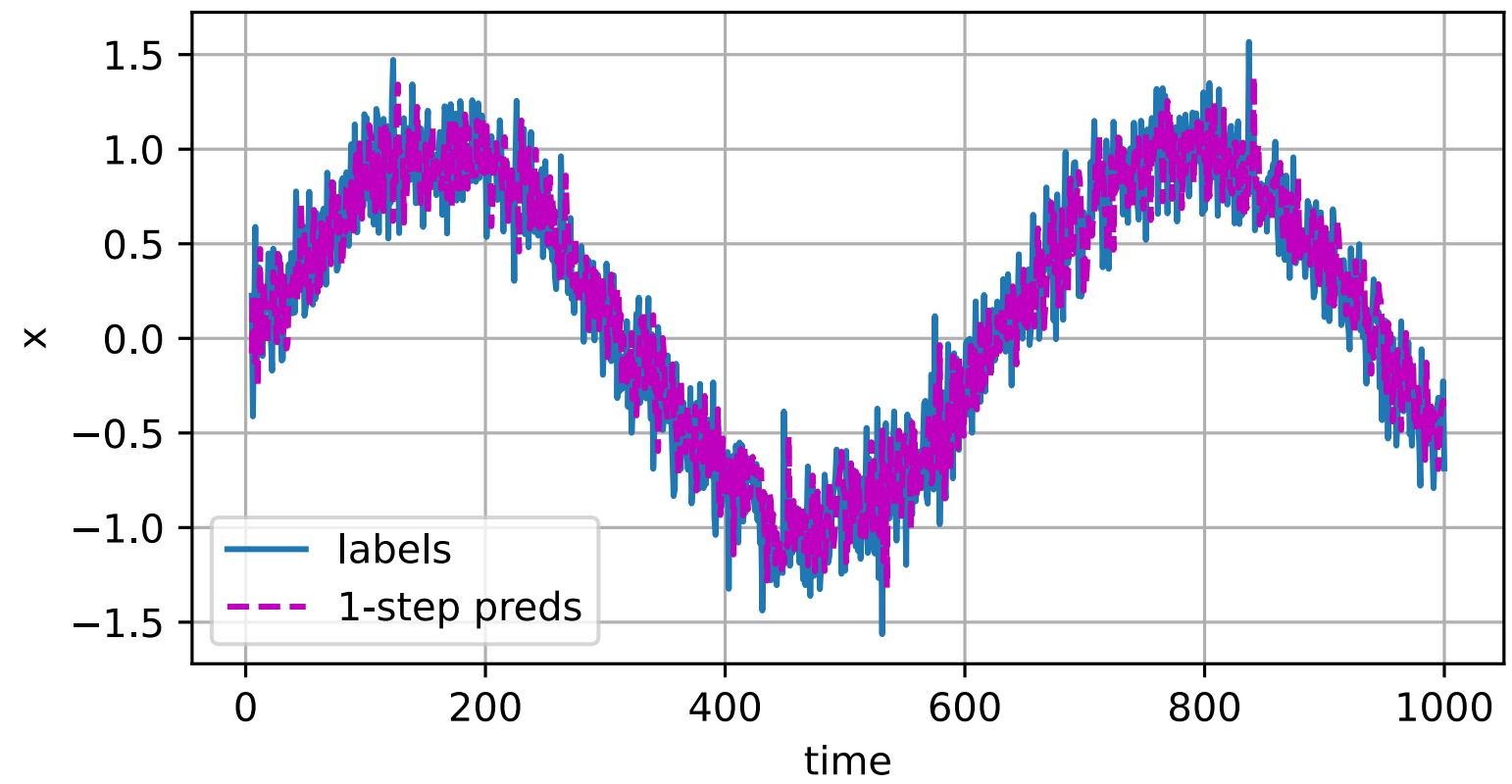

These predictions look good, even near the end at $t = 1000$.

But what if we only observed sequence data up until time step 604 (`n_train + tau`) and wished to make predictions several steps into the future? Unfortunately, we cannot directly compute the one-step-ahead prediction for time step 609, because we do not know the corresponding inputs, having seen only up to $x_{604}$. We can address this problem by plugging in our earlier predictions as inputs to our model for making subsequent predictions, projecting forward, one step at a time, until reaching the desired time step:

$$\begin{aligned}
\hat{x}_{605} &= f(x_{601}, x_{602}, x_{603}, x_{604}), \\
\hat{x}_{606} &= f(x_{602}, x_{603}, x_{604}, \hat{x}_{605}), \\
\hat{x}_{607} &= f(x_{603}, x_{604}, \hat{x}_{605}, \hat{x}_{606}), \\
\hat{x}_{608} &= f(x_{604}, \hat{x}_{605}, \hat{x}_{606}, \hat{x}_{607}), \\
\hat{x}_{609} &= f(\hat{x}_{605}, \hat{x}_{606}, \hat{x}_{607}, \hat{x}_{608}), \\
&\vdots
\end{aligned} \tag{9.1.6}$$

Generally, for an observed sequence $x_1, \ldots, x_t$, its predicted output $\hat{x}_{t+k}$ at time step $t + k$ is called the $k$*-step-ahead prediction*. Since we have observed up to $x_{604}$, its $k$-step-ahead prediction is $\hat{x}_{604+k}$. In other words, we will have to keep on using our own predictions to make multistep-ahead predictions. Let's see how well this goes.

```
multistep_preds = torch.zeros(data.T)
multistep_preds[:] = data.x
for i in range(data.num_train + data.tau, data.T):
    multistep_preds[i] = model(
        multistep_preds[i - data.tau:i].reshape((1, -1)))
multistep_preds = multistep_preds.detach().numpy()
```

```
d2l.plot([data.time[data.tau:], data.time[data.num_train+data.tau:]],
         [onestep_preds, multistep_preds[data.num_train+data.tau:]], 'time',
         'x', legend=['1-step preds', 'multistep preds'], figsize=(6, 3))
```

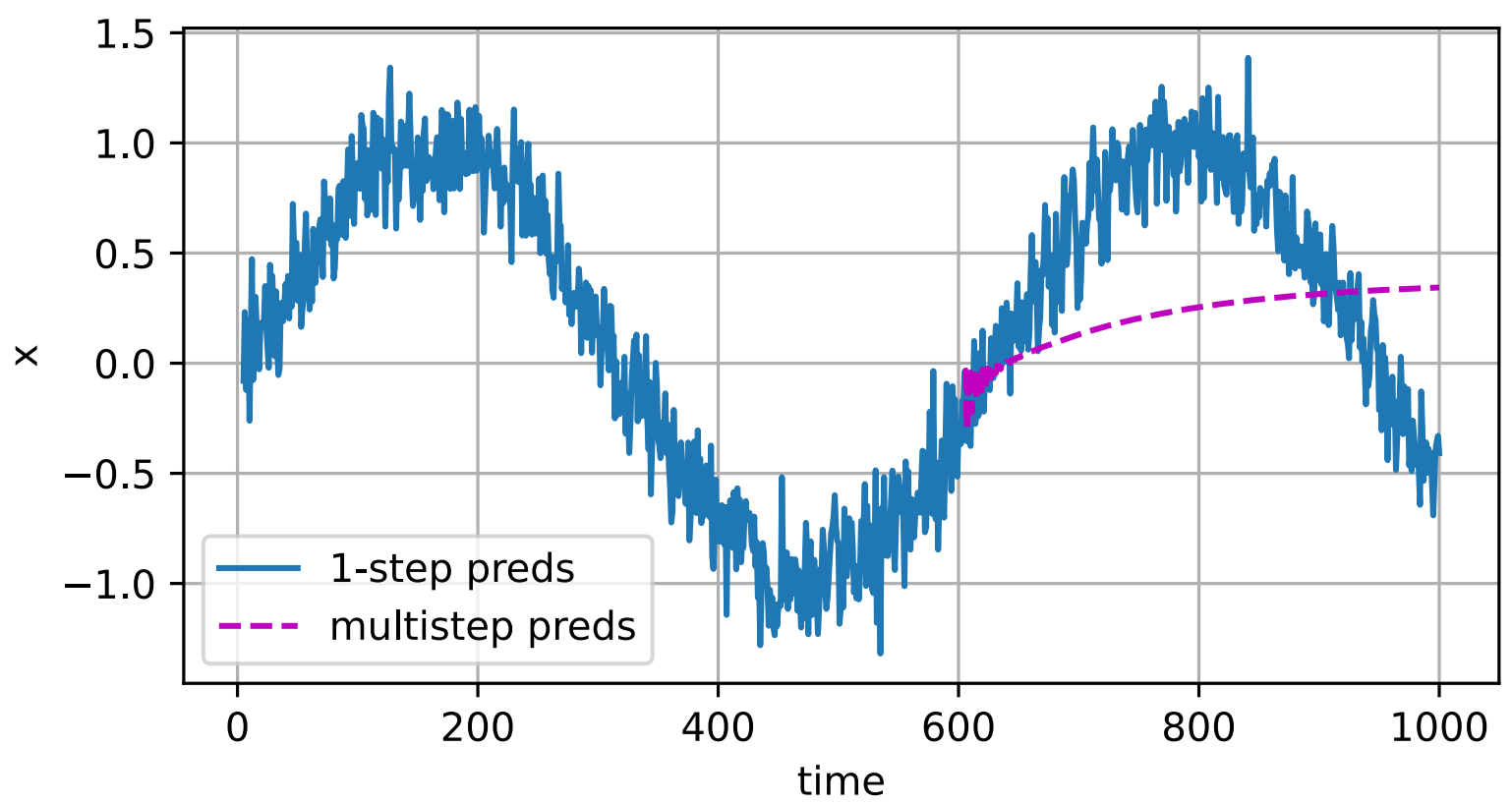

Unfortunately, in this case we fail spectacularly. The predictions decay to a constant pretty quickly after a few steps. Why did the algorithm perform so much worse when predicting further into the future? Ultimately, this is down to the fact that errors build up. Let's say that after step 1 we have some error $\epsilon_1 = \bar{\epsilon}$. Now the *input* for step 2 is perturbed by $\epsilon_1$, hence we suffer some error in the order of $\epsilon_2 = \bar{\epsilon} + c\epsilon_1$ for some constant $c$, and so on. The predictions can diverge rapidly from the true observations. You may already be familiar with this common phenomenon. For instance, weather forecasts for the next 24 hours tend to be pretty accurate but beyond that, accuracy declines rapidly. We will discuss methods for improving this throughout this chapter and beyond.

Let's take a closer look at the difficulties in $k$-step-ahead predictions by computing predictions on the entire sequence for $k = 1, 4, 16, 64$.

```
def k_step_pred(k):
    features = []
    for i in range(data.tau):
        features.append(data.x[i : i+data.T-data.tau-k+1])
    # The (i+tau)-th element stores the (i+1)-step-ahead predictions
    for i in range(k):
        preds = model(torch.stack(features[i : i+data.tau], 1))
        features.append(preds.reshape(-1))
    return features[data.tau:]
```

```
steps = (1, 4, 16, 64)
preds = k_step_pred(steps[-1])
d2l.plot(data.time[data.tau+steps[-1]-1:],
         [preds[k - 1].detach().numpy() for k in steps], 'time', 'x',
         legend=[f'{k}-step preds' for k in steps], figsize=(6, 3))
```

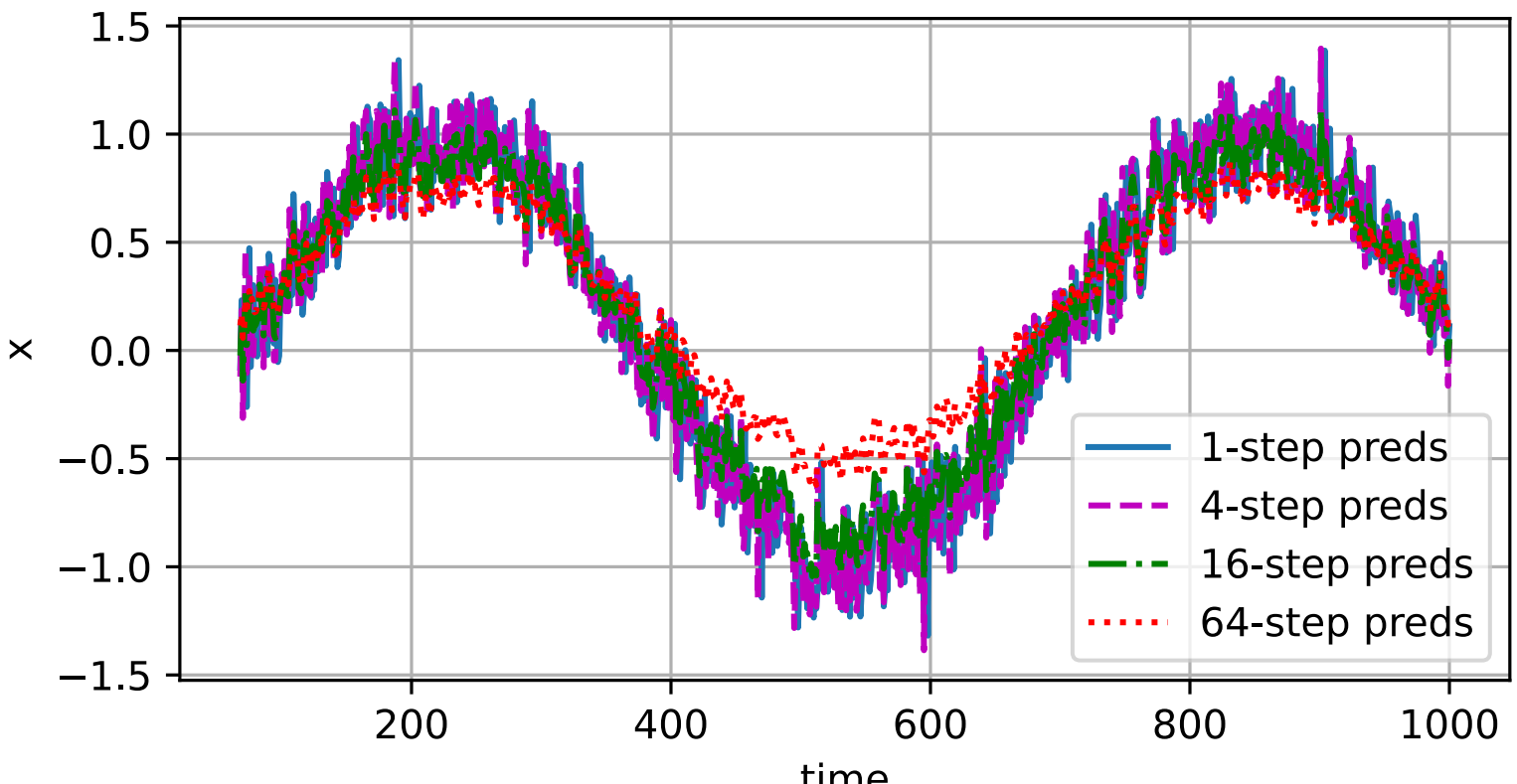

This clearly illustrates how the quality of the prediction changes as we try to predict further into the future. While the 4-step-ahead predictions still look good, anything beyond that is almost useless.

### 9.1.5 Summary

There is quite a difference in difficulty between interpolation and extrapolation. Consequently, if you have a sequence, always respect the temporal order of the data when training, i.e., never train on future data. Given this kind of data, sequence models require specialized statistical tools for estimation. Two popular choices are autoregressive models and latent-variable autoregressive models. For causal models (e.g., time going forward), estimating the forward direction is typically a lot easier than the reverse direction. For an observed sequence up to time step $t$, its predicted output at time step $t + k$ is the $k$-*step-ahead prediction*. As we predict further in time by increasing $k$, the errors accumulate and the quality of the prediction degrades, often dramatically.

### 9.1.6 Exercises

1. Improve the model in the experiment of this section.
    1. Incorporate more than the past four observations? How many do you really need?
    2. How many past observations would you need if there was no noise? Hint: you can write sin and cos as a differential equation.
    3. Can you incorporate older observations while keeping the total number of features constant? Does this improve accuracy? Why?
    4. Change the neural network architecture and evaluate the performance. You may train the new model with more epochs. What do you observe?
2. An investor wants to find a good security to buy. They look at past returns to decide which one is likely to do well. What could possibly go wrong with this strategy?
3. Does causality also apply to text? To which extent?
4. Give an example for when a latent autoregressive model might be needed to capture the dynamic of the data.

Discussions[136].

## 9.2 Converting Raw Text into Sequence Data

Throughout this book, we will often work with text data represented as sequences of words, characters, or word pieces. To get going, we will need some basic tools for converting raw text into sequences of the appropriate form. Typical preprocessing pipelines execute the following steps:

1. Load text as strings into memory.
2. Split the strings into tokens (e.g., words or characters).
3. Build a vocabulary dictionary to associate each vocabulary element with a numerical index.
4. Convert the text into sequences of numerical indices.

```
import collections
import random
import re
import torch
from d2l import torch as d2l
```

### 9.2.1 Reading the Dataset

Here, we will work with H. G. Wells' The Time Machine [137], a book containing just over 30,000 words. While real applications will typically involve significantly larger datasets, this is sufficient to demonstrate the preprocessing pipeline. The following `_download` method reads the raw text into a string.

```
class TimeMachine(d2l.DataModule): #@save
    """The Time Machine dataset."""
    def _download(self):
        fname = d2l.download(d2l.DATA_URL + 'timemachine.txt', self.root,
                             '090b5e7e70c295757f55df93cb0a180b9691891a')
        with open(fname) as f:
            return f.read()

data = TimeMachine()
raw_text = data._download()
raw_text[:60]
```

```
'The Time Machine, by H. G. Wells [1898]nnnnnInnnThe Time Tra'
```

For simplicity, we ignore punctuation and capitalization when preprocessing the raw text.

```
@d2l.add_to_class(TimeMachine)  #@save
def _preprocess(self, text):
    return re.sub('[^A-Za-z]+', ' ', text).lower()

text = data._preprocess(raw_text)
text[:60]
```

```
'the time machine by h g wells i the time traveller for so it'
```

### 9.2.2 Tokenization

*Tokens* are the atomic (indivisible) units of text. Each time step corresponds to 1 token, but what precisely constitutes a token is a design choice. For example, we could represent the sentence "Baby needs a new pair of shoes" as a sequence of 7 words, where the set of all words comprise a large vocabulary (typically tens or hundreds of thousands of words). Or we would represent the same sentence as a much longer sequence of 30 characters, using a much smaller vocabulary (there are only 256 distinct ASCII characters). Below, we tokenize our preprocessed text into a sequence of characters.

```
@d2l.add_to_class(TimeMachine)  #@save
def _tokenize(self, text):
    return list(text)

tokens = data._tokenize(text)
','.join(tokens[:30])
```

```
't,h,e, ,t,i,m,e, ,m,a,c,h,i,n,e, ,b,y, ,h, ,g, ,w,e,l,l,s, '
```

### 9.2.3 Vocabulary

These tokens are still strings. However, the inputs to our models must ultimately consist of numerical inputs. Next, we introduce a class for constructing *vocabularies*, i.e., objects that associate each distinct token value with a unique index. First, we determine the set of unique tokens in our training *corpus*. We then assign a numerical index to each unique token. Rare vocabulary elements are often dropped for convenience. Whenever we encounter a token at training or test time that had not been previously seen or was dropped from the vocabulary, we represent it by a special “<unk>” token, signifying that this is an *unknown* value.

```
class Vocab:  #@save
    """Vocabulary for text."""
    def __init__(self, tokens=[], min_freq=0, reserved_tokens=[]):
        # Flatten a 2D list if needed
        if tokens and isinstance(tokens[0], list):
            tokens = [token for line in tokens for token in line]
        # Count token frequencies
        counter = collections.Counter(tokens)
        self.token_freqs = sorted(counter.items(), key=lambda x: x[1],
                                  reverse=True)
        # The list of unique tokens
        self.idx_to_token = list(sorted(set(['<unk>'] + reserved_tokens + [
            token for token, freq in self.token_freqs if freq >= min_freq])))
        self.token_to_idx = {token: idx
                             for idx, token in enumerate(self.idx_to_token)}

    def __len__(self):
        return len(self.idx_to_token)

    def __getitem__(self, tokens):
        if not isinstance(tokens, (list, tuple)):
            return self.token_to_idx.get(tokens, self.unk)
        return [self.__getitem__(token) for token in tokens]

    def to_tokens(self, indices):
        if hasattr(indices, '__len__') and len(indices) > 1:
            return [self.idx_to_token[int(index)] for index in indices]
        return self.idx_to_token[indices]
```

```
    @property
    def unk(self):  # Index for the unknown token
        return self.token_to_idx['<unk>']
```

We now construct a vocabulary for our dataset, converting the sequence of strings into a list of numerical indices. Note that we have not lost any information and can easily convert our dataset back to its original (string) representation.

```
vocab = Vocab(tokens)
indices = vocab[tokens[:10]]
print('indices:', indices)
print('words:', vocab.to_tokens(indices))
```

```
indices: [21, 9, 6, 0, 21, 10, 14, 6, 0, 14]
words: ['t', 'h', 'e', ' ', 't', 'i', 'm', 'e', ' ', 'm']
```

## 9.2.4 Putting It All Together

Using the above classes and methods, we package everything into the following `build` method of the `TimeMachine` class, which returns `corpus`, a list of token indices, and `vocab`, the vocabulary of *The Time Machine* corpus. The modifications we did here are: (i) we tokenize text into characters, not words, to simplify the training in later sections; (ii) `corpus` is a single list, not a list of token lists, since each text line in *The Time Machine* dataset is not necessarily a sentence or paragraph.

```
@d2l.add_to_class(TimeMachine)  #@save
def build(self, raw_text, vocab=None):
    tokens = self._tokenize(self._preprocess(raw_text))
    if vocab is None: vocab = Vocab(tokens)
    corpus = [vocab[token] for token in tokens]
    return corpus, vocab

corpus, vocab = data.build(raw_text)
len(corpus), len(vocab)
```

```
(173428, 28)
```

## 9.2.5 Exploratory Language Statistics

Using the real corpus and the `Vocab` class defined over words, we can inspect basic statistics concerning word use in our corpus. Below, we construct a vocabulary from words used in *The Time Machine* and print the ten most frequently occurring of them.

```
words = text.split()
vocab = Vocab(words)
vocab.token_freqs[:10]
```

```
[('the', 2261),
 ('i', 1267),
 ('and', 1245),
 ('of', 1155),
 ('a', 816),
 ('to', 695),
 ('was', 552),
 ('in', 541),
 ('that', 443),
 ('my', 440)]
```

Note that the ten most frequent words are not all that descriptive. You might even imagine that we might see a very similar list if we had chosen any book at random. Articles like "the" and "a", pronouns like "i" and "my", and prepositions like "of", "to", and "in" occur often because they serve common syntactic roles. Such words that are common but not particularly descriptive are often called *stop words* and, in previous generations of text classifiers based on so-called bag-of-words representations, they were most often filtered out. However, they carry meaning and it is not necessary to filter them out when working with modern RNN- and Transformer-based neural models. If you look further down the list, you will notice that word frequency decays quickly. The $10^{\text{th}}$ most frequent word is less than $1/5$ as common as the most popular. Word frequency tends to follow a power law distribution (specifically the Zipfian) as we go down the ranks. To get a better idea, we plot the figure of the word frequency.

```
freqs = [freq for token, freq in vocab.token_freqs]
d2l.plot(freqs, xlabel='token: x', ylabel='frequency: n(x)',
         xscale='log', yscale='log')
```

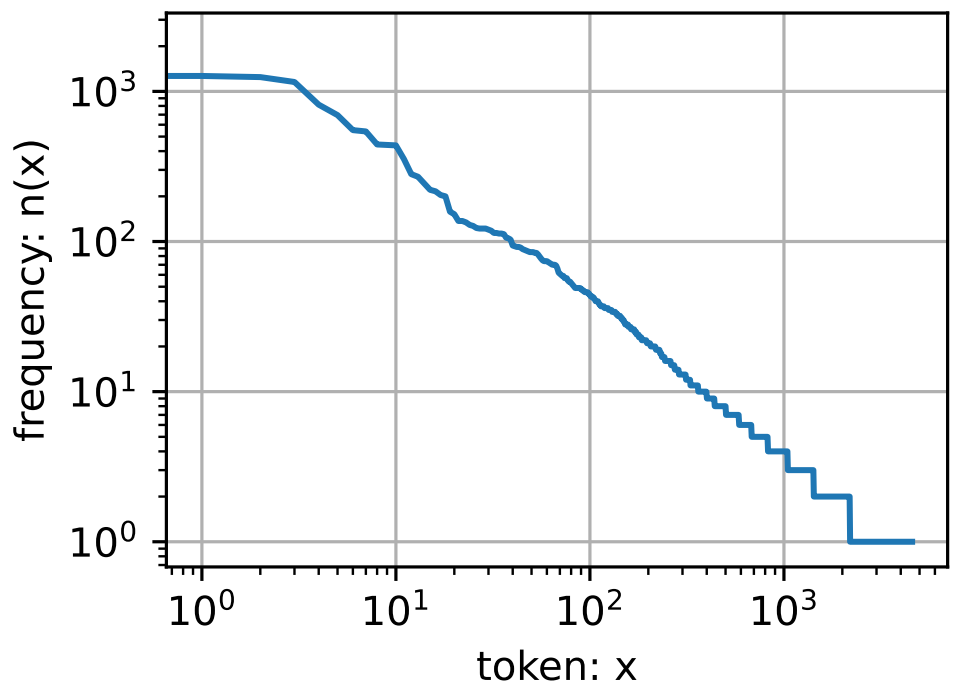

After dealing with the first few words as exceptions, all the remaining words roughly follow a straight line on a log–log plot. This phenomenon is captured by *Zipf's law*, which states that the frequency $n_i$ of the $i^{\text{th}}$ most frequent word is:

$$n_i \propto \frac{1}{i^\alpha}, \tag{9.2.1}$$

which is equivalent to

$$\log n_i = -\alpha \log i + c, \tag{9.2.2}$$

where $\alpha$ is the exponent that characterizes the distribution and $c$ is a constant. This should already give us pause for thought if we want to model words by counting statistics. After all, we will significantly overestimate the frequency of the tail, also known as the infrequent words. But what about the other word combinations, such as two consecutive words (bigrams), three consecutive words (trigrams), and beyond? Let's see whether the bigram frequency behaves in the same manner as the single word (unigram) frequency.

```
bigram_tokens = ['--'.join(pair) for pair in zip(words[:-1], words[1:])]
bigram_vocab = Vocab(bigram_tokens)
bigram_vocab.token_freqs[:10]
```

```
[('of--the', 309),
 ('in--the', 169),
 ('i--had', 130),
 ('i--was', 112),
 ('and--the', 109),
 ('the--time', 102),
 ('it--was', 99),
 ('to--the', 85),
 ('as--i', 78),
 ('of--a', 73)]
```

One thing is notable here. Out of the ten most frequent word pairs, nine are composed of both stop words and only one is relevant to the actual book—"the time". Furthermore, let's see whether the trigram frequency behaves in the same manner.

```
trigram_tokens = ['--'.join(triple) for triple in zip(
    words[:-2], words[1:-1], words[2:])]
trigram_vocab = Vocab(trigram_tokens)
trigram_vocab.token_freqs[:10]
```

```
[('the--time--traveller', 59),
 ('the--time--machine', 30),
 ('the--medical--man', 24),
 ('it--seemed--to', 16),
 ('it--was--a', 15),
 ('here--and--there', 15),
 ('seemed--to--me', 14),
 ('i--did--not', 14),
 ('i--saw--the', 13),
 ('i--began--to', 13)]
```

Now, let's visualize the token frequency among these three models: unigrams, bigrams, and trigrams.

```
bigram_freqs = [freq for token, freq in bigram_vocab.token_freqs]
trigram_freqs = [freq for token, freq in trigram_vocab.token_freqs]
d2l.plot([freqs, bigram_freqs, trigram_freqs], xlabel='token: x',
         ylabel='frequency: n(x)', xscale='log', yscale='log',
         legend=['unigram', 'bigram', 'trigram'])
```

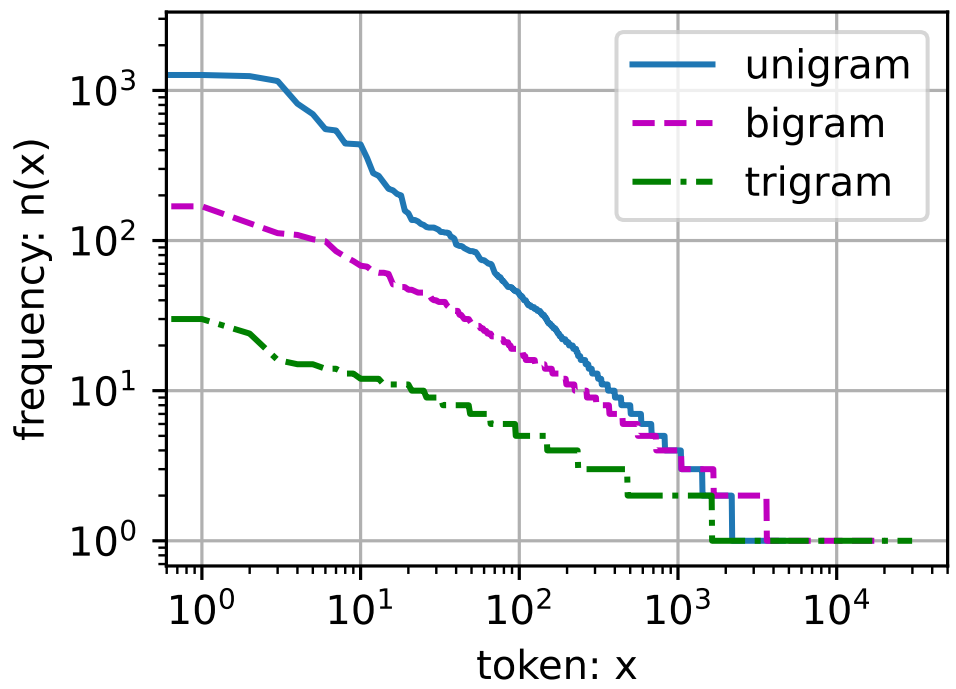

This figure is quite exciting. First, beyond unigram words, sequences of words also appear to be following Zipf's law, albeit with a smaller exponent $\alpha$ in (9.2.1), depending on the sequence length. Second, the number of distinct $n$-grams is not that large. This gives us hope that there is quite a lot of structure in language. Third, many $n$-grams occur very rarely. This makes certain methods unsuitable for language modeling and motivates the use of deep learning models. We will discuss this in the next section.

## 9.2.6 Summary

Text is among the most common forms of sequence data encountered in deep learning. Common choices for what constitutes a token are characters, words, and word pieces. To preprocess text, we usually (i) split text into tokens; (ii) build a vocabulary to map token strings to numerical indices; and (iii) convert text data into token indices for models to manipulate. In practice, the frequency of words tends to follow Zipf's law. This is true not just for individual words (unigrams), but also for $n$-grams.

## 9.2.7 Exercises

1. In the experiment of this section, tokenize text into words and vary the `min_freq` argument value of the `Vocab` instance. Qualitatively characterize how changes in `min_freq` impact the size of the resulting vocabulary.
2. Estimate the exponent of Zipfian distribution for unigrams, bigrams, and trigrams in this corpus.

3. Find some other sources of data (download a standard machine learning dataset, pick another public domain book, scrape a website, etc). For each, tokenize the data at both the word and character levels. How do the vocabulary sizes compare with *The Time Machine* corpus at equivalent values of `min_freq`. Estimate the exponent of the Zipfian distribution corresponding to the unigram and bigram distributions for these corpora. How do they compare with the values that you observed for *The Time Machine* corpus?

Discussions[138].

# 9.3 Language Models

In Section 9.2, we saw how to map text sequences into tokens, where these tokens can be viewed as a sequence of discrete observations such as words or characters. Assume that the tokens in a text sequence of length $T$ are in turn $x_1, x_2, \ldots, x_T$. The goal of *language models* is to estimate the joint probability of the whole sequence:

$$P(x_1, x_2, \ldots, x_T), \tag{9.3.1}$$

where statistical tools in Section 9.1 can be applied.

Language models are incredibly useful. For instance, an ideal language model should generate natural text on its own, simply by drawing one token at a time $x_t \sim P(x_t \mid x_{t-1}, \ldots, x_1)$. Quite unlike the monkey using a typewriter, all text emerging from such a model would pass as natural language, e.g., English text. Furthermore, it would be sufficient for generating a meaningful dialog, simply by conditioning the text on previous dialog fragments. Clearly we are still very far from designing such a system, since it would need to *understand* the text rather than just generate grammatically sensible content.

Nonetheless, language models are of great service even in their limited form. For instance, the phrases "to recognize speech" and "to wreck a nice beach" sound very similar. This can cause ambiguity in speech recognition, which is easily resolved through a language model that rejects the second translation as outlandish. Likewise, in a document summarization algorithm it is worthwhile knowing that "dog bites man" is much more frequent than "man bites dog", or that "I want to eat grandma" is a rather disturbing statement, whereas "I want to eat, grandma" is much more benign.

```
import torch
from d2l import torch as d2l
```

## 9.3.1 Learning Language Models

The obvious question is how we should model a document, or even a sequence of tokens. Suppose that we tokenize text data at the word level. Let's start by applying basic probability rules:

$$P(x_1, x_2, \ldots, x_T) = \prod_{t=1}^{T} P(x_t \mid x_1, \ldots, x_{t-1}). \tag{9.3.2}$$

For example, the probability of a text sequence containing four words would be given as:

$$\begin{aligned} &P(\text{deep}, \text{learning}, \text{is}, \text{fun}) \\ =&P(\text{deep})P(\text{learning} \mid \text{deep})P(\text{is} \mid \text{deep}, \text{learning})P(\text{fun} \mid \text{deep}, \text{learning}, \text{is}). \end{aligned} \tag{9.3.3}$$

## Markov Models and $n$-grams

Among those sequence model analyses in Section 9.1, let's apply Markov models to language modeling. A distribution over sequences satisfies the Markov property of first order if $P(x_{t+1} \mid x_t, \ldots, x_1) = P(x_{t+1} \mid x_t)$. Higher orders correspond to longer dependencies. This leads to a number of approximations that we could apply to model a sequence:

$$\begin{aligned} P(x_1, x_2, x_3, x_4) &= P(x_1) P(x_2) P(x_3) P(x_4),\\ P(x_1, x_2, x_3, x_4) &= P(x_1) P(x_2 \mid x_1) P(x_3 \mid x_2) P(x_4 \mid x_3),\\ P(x_1, x_2, x_3, x_4) &= P(x_1) P(x_2 \mid x_1) P(x_3 \mid x_1, x_2) P(x_4 \mid x_2, x_3). \end{aligned} \tag{9.3.4}$$

The probability formulae that involve one, two, and three variables are typically referred to as *unigram*, *bigram*, and *trigram* models, respectively. In order to compute the language model, we need to calculate the probability of words and the conditional probability of a word given the previous few words. Note that such probabilities are language model parameters.

## Word Frequency

Here, we assume that the training dataset is a large text corpus, such as all Wikipedia entries, Project Gutenberg[139], and all text posted on the web. The probability of words can be calculated from the relative word frequency of a given word in the training dataset. For example, the estimate $\hat{P}(\text{deep})$ can be calculated as the probability of any sentence starting with the word "deep". A slightly less accurate approach would be to count all occurrences of the word "deep" and divide it by the total number of words in the corpus. This works fairly well, particularly for frequent words. Moving on, we could attempt to estimate

$$\hat{P}(\text{learning} \mid \text{deep}) = \frac{n(\text{deep, learning})}{n(\text{deep})}, \tag{9.3.5}$$

where $n(x)$ and $n(x, x')$ are the number of occurrences of singletons and consecutive word pairs, respectively. Unfortunately, estimating the probability of a word pair is somewhat more difficult, since the occurrences of "deep learning" are a lot less frequent. In particular, for some unusual word combinations it may be tricky to find enough occurrences to get accurate estimates. As suggested by the empirical results in Section 9.2.5, things take a turn for the worse for three-word combinations and beyond. There will be many plausible three-word combinations that we likely will not see in our dataset. Unless we provide some solution to assign such word combinations a nonzero count, we will not be able to use them in a language model. If the dataset is small or if the words are very rare, we might not find even a single one of them.

### Laplace Smoothing

A common strategy is to perform some form of *Laplace smoothing*. The solution is to add a small constant to all counts. Denote by $n$ the total number of words in the training set and $m$ the number of unique words. This solution helps with singletons, e.g., via

$$\begin{aligned}
\hat{P}(x) & = \frac{n(x) + \epsilon_1/m}{n + \epsilon_1}, \\
\hat{P}(x' \mid x) & = \frac{n(x, x') + \epsilon_2 \hat{P}(x')}{n(x) + \epsilon_2}, \\
\hat{P}(x'' \mid x, x') & = \frac{n(x, x', x'') + \epsilon_3 \hat{P}(x'')}{n(x, x') + \epsilon_3}.
\end{aligned} \tag{9.3.6}$$

Here $\epsilon_1, \epsilon_2$, and $\epsilon_3$ are hyperparameters. Take $\epsilon_1$ as an example: when $\epsilon_1 = 0$, no smoothing is applied; when $\epsilon_1$ approaches positive infinity, $\hat{P}(x)$ approaches the uniform probability $1/m$. The above is a rather primitive variant of what other techniques can accomplish (Wood *et al.*, 2011).

Unfortunately, models like this get unwieldy rather quickly for the following reasons. First, as discussed in Section 9.2.5, many $n$-grams occur very rarely, making Laplace smoothing rather unsuitable for language modeling. Second, we need to store all counts. Third, this entirely ignores the meaning of the words. For instance, "cat" and "feline" should occur in related contexts. It is quite difficult to adjust such models to additional contexts, whereas, deep learning based language models are well suited to take this into account. Last, long word sequences are almost certain to be novel, hence a model that simply counts the frequency of previously seen word sequences is bound to perform poorly there. Therefore, we focus on using neural networks for language modeling in the rest of the chapter.

## 9.3.2 Perplexity

Next, let's discuss about how to measure the quality of the language model, which we will then use to evaluate our models in the subsequent sections. One way is to check how surprising the text is. A good language model is able to predict, with high accuracy, the tokens that come next. Consider the following continuations of the phrase "It is raining", as proposed by different language models:

1. "It is raining outside"
2. "It is raining banana tree"

3. "It is raining piouw;kcj pwepoiut"

In terms of quality, Example 1 is clearly the best. The words are sensible and logically coherent. While it might not quite accurately reflect which word follows semantically ("in San Francisco" and "in winter" would have been perfectly reasonable extensions), the model is able to capture which kind of word follows. Example 2 is considerably worse by producing a nonsensical extension. Nonetheless, at least the model has learned how to spell words and some degree of correlation between words. Last, Example 3 indicates a poorly trained model that does not fit data properly.

We might measure the quality of the model by computing the likelihood of the sequence. Unfortunately this is a number that is hard to understand and difficult to compare. After all, shorter sequences are much more likely to occur than the longer ones, hence evaluating the model on Tolstoy's magnum opus *War and Peace* will inevitably produce a much smaller likelihood than, say, on Saint-Exupery's novella *The Little Prince*. What is missing is the equivalent of an average.

Information theory comes handy here. We defined entropy, surprisal, and cross-entropy when we introduced the softmax regression (Section 4.1.3). If we want to compress text, we can ask about predicting the next token given the current set of tokens. A better language model should allow us to predict the next token more accurately. Thus, it should allow us to spend fewer bits in compressing the sequence. So we can measure it by the cross-entropy loss averaged over all the $n$ tokens of a sequence:

$$\frac{1}{n} \sum_{t=1}^{n} -\log P(x_t \mid x_{t-1}, \ldots, x_1), \tag{9.3.7}$$

where $P$ is given by a language model and $x_t$ is the actual token observed at time step $t$ from the sequence. This makes the performance on documents of different lengths comparable. For historical reasons, scientists in natural language processing prefer to use a quantity called *perplexity*. In a nutshell, it is the exponential of (9.3.7):

$$\exp\left(-\frac{1}{n} \sum_{t=1}^{n} \log P(x_t \mid x_{t-1}, \ldots, x_1)\right). \tag{9.3.8}$$

Perplexity can be best understood as the reciprocal of the geometric mean of the number of real choices that we have when deciding which token to pick next. Let's look at a number of cases:

- In the best case scenario, the model always perfectly estimates the probability of the target token as 1. In this case the perplexity of the model is 1.
- In the worst case scenario, the model always predicts the probability of the target token as 0. In this situation, the perplexity is positive infinity.
- At the baseline, the model predicts a uniform distribution over all the available tokens of the vocabulary. In this case, the perplexity equals the number of unique tokens of the vocabulary. In fact, if we were to store the sequence without any compression, this would be the best we could do for encoding it. Hence, this provides a nontrivial upper bound that any useful model must beat.

### 9.3.3 Partitioning Sequences

We will design language models using neural networks and use perplexity to evaluate how good the model is at predicting the next token given the current set of tokens in text sequences. Before introducing the model, let's assume that it processes a minibatch of sequences with predefined length at a time. Now the question is how to read minibatches of input sequences and target sequences at random.

Suppose that the dataset takes the form of a sequence of $T$ token indices in `corpus`. We will partition it into subsequences, where each subsequence has $n$ tokens (time steps). To iterate over (almost) all the tokens of the entire dataset for each epoch and obtain all possible length-$n$ subsequences, we can introduce randomness. More concretely, at the beginning of each epoch, discard the first $d$ tokens, where $d \in [0, n)$ is uniformly sampled at random. The rest of the sequence is then partitioned into $m = \lfloor (T-d)/n \rfloor$ subsequences. Denote by $\mathbf{x}_t = [x_t, \ldots, x_{t+n-1}]$ the length-$n$ subsequence starting from token $x_t$ at time step $t$. The resulting $m$ partitioned subsequences are $\mathbf{x}_d, \mathbf{x}_{d+n}, \ldots, \mathbf{x}_{d+n(m-1)}$. Each subsequence will be used as an input sequence into the language model.

For language modeling, the goal is to predict the next token based on the tokens we have seen so far; hence the targets (labels) are the original sequence, shifted by one token. The target sequence for any input sequence $\mathbf{x}_t$ is $\mathbf{x}_{t+1}$ with length $n$.

Input sequences: the time machine by h g wells

Target sequences: the time machine by h g wells

Fig. 9.3.1 Obtaining five pairs of input sequences and target sequences from partitioned length-5 subsequences.

Fig. 9.3.1 shows an example of obtaining five pairs of input sequences and target sequences with $n = 5$ and $d = 2$.

```
@d2l.add_to_class(d2l.TimeMachine)  #@save
def __init__(self, batch_size, num_steps, num_train=10000, num_val=5000):
    super(d2l.TimeMachine, self).__init__()
    self.save_hyperparameters()
    corpus, self.vocab = self.build(self._download())
    array = torch.tensor([corpus[i:i+num_steps+1]
                        for i in range(len(corpus)-num_steps)])
    self.X, self.Y = array[:,:-1], array[:,1:]
```

To train language models, we will randomly sample pairs of input sequences and target sequences in minibatches. The following data loader randomly generates a minibatch from the dataset each time. The argument `batch_size` specifies the number of subsequence examples in each minibatch and `num_steps` is the subsequence length in tokens.

```
@d2l.add_to_class(d2l.TimeMachine)  #@save
def get_dataloader(self, train):
```

```
    idx = slice(0, self.num_train) if train else slice(
        self.num_train, self.num_train + self.num_val)
    return self.get_tensorloader([self.X, self.Y], train, idx)
```

As we can see in the following, a minibatch of target sequences can be obtained by shifting the input sequences by one token.

```
data = d2l.TimeMachine(batch_size=2, num_steps=10)
for X, Y in data.train_dataloader():
    print('X:', X, '\nY:', Y)
    break
```

```
X: tensor([[10, 15, 12,  0, 26, 16, 22,  0, 24, 16],
        [ 9,  0,  2,  0, 22, 15, 10,  7, 16, 19]])
Y: tensor([[15, 12,  0, 26, 16, 22,  0, 24, 16, 22],
        [ 0,  2,  0, 22, 15, 10,  7, 16, 19, 14]])
```

### 9.3.4 Summary and Discussion

Language models estimate the joint probability of a text sequence. For long sequences, $n$-grams provide a convenient model by truncating the dependence. However, there is a lot of structure but not enough frequency to deal efficiently with infrequent word combinations via Laplace smoothing. Thus, we will focus on neural language modeling in subsequent sections. To train language models, we can randomly sample pairs of input sequences and target sequences in minibatches. After training, we will use perplexity to measure the language model quality.

Language models can be scaled up with increased data size, model size, and amount in training compute. Large language models can perform desired tasks by predicting output text given input text instructions. As we will discuss later (e.g., Section 11.9), at the present moment large language models form the basis of state-of-the-art systems across diverse tasks.

### 9.3.5 Exercises

1. Suppose there are 100,000 words in the training dataset. How much word frequency and multi-word adjacent frequency does a four-gram need to store?
2. How would you model a dialogue?
3. What other methods can you think of for reading long sequence data?
4. Consider our method for discarding a uniformly random number of the first few tokens at the beginning of each epoch.
    1. Does it really lead to a perfectly uniform distribution over the sequences on the document?

2. What would you have to do to make things even more uniform?

5. If we want a sequence example to be a complete sentence, what kind of problem does this introduce in minibatch sampling? How can we fix it?

Discussions[140].

## 9.4 Recurrent Neural Networks

In Section 9.3 we described Markov models and $n$-grams for language modeling, where the conditional probability of token $x_t$ at time step $t$ only depends on the $n-1$ previous tokens. If we want to incorporate the possible effect of tokens earlier than time step $t-(n-1)$ on $x_t$, we need to increase $n$. However, the number of model parameters would also increase exponentially with it, as we need to store $|\mathcal{V}|^n$ numbers for a vocabulary set $\mathcal{V}$. Hence, rather than modeling $P(x_t \mid x_{t-1}, \ldots, x_{t-n+1})$ it is preferable to use a latent variable model,

$$P(x_t \mid x_{t-1}, \ldots, x_1) \approx P(x_t \mid h_{t-1}), \tag{9.4.1}$$

where $h_{t-1}$ is a *hidden state* that stores the sequence information up to time step $t-1$. In general, the hidden state at any time step $t$ could be computed based on both the current input $x_t$ and the previous hidden state $h_{t-1}$:

$$h_t = f(x_t, h_{t-1}). \tag{9.4.2}$$

For a sufficiently powerful function $f$ in (9.4.2), the latent variable model is not an approximation. After all, $h_t$ may simply store all the data it has observed so far. However, it could potentially make both computation and storage expensive.

Recall that we have discussed hidden layers with hidden units in Chapter 5. It is noteworthy that hidden layers and hidden states refer to two very different concepts. Hidden layers are, as explained, layers that are hidden from view on the path from input to output. Hidden states are technically speaking *inputs* to whatever we do at a given step, and they can only be computed by looking at data at previous time steps.

*Recurrent neural networks* (RNNs) are neural networks with hidden states. Before introducing the RNN model, we first revisit the MLP model introduced in Section 5.1.

```
import torch
from d2l import torch as d2l
```

### 9.4.1 Neural Networks without Hidden States

Let's take a look at an MLP with a single hidden layer. Let the hidden layer's activation function be $\phi$. Given a minibatch of examples $\mathbf{X} \in \mathbb{R}^{n \times d}$ with batch size $n$ and $d$ inputs, the hidden layer output $\mathbf{H} \in \mathbb{R}^{n\times h}$ is calculated as

$$\mathbf{H} = \phi(\mathbf{X}\mathbf{W}_{\mathrm{xh}} + \mathbf{b}_{\mathrm{h}}). \tag{9.4.3}$$

In (9.4.3), we have the weight parameter $\mathbf{W}_{\mathrm{xh}} \in \mathbb{R}^{d\times h}$, the bias parameter $\mathbf{b}_{\mathrm{h}} \in \mathbb{R}^{1\times h}$, and the number of hidden units $h$, for the hidden layer. So armed, we apply broadcasting (see Section 2.1.4) during the summation. Next, the hidden layer output $\mathbf{H}$ is used as input of the output layer, which is given by

$$\mathbf{O} = \mathbf{H}\mathbf{W}_{\mathrm{hq}} + \mathbf{b}_{\mathrm{q}}, \tag{9.4.4}$$

where $\mathbf{O} \in \mathbb{R}^{n\times q}$ is the output variable, $\mathbf{W}_{\mathrm{hq}} \in \mathbb{R}^{h\times q}$ is the weight parameter, and $\mathbf{b}_{\mathrm{q}} \in \mathbb{R}^{1\times q}$ is the bias parameter of the output layer. If it is a classification problem, we can use $\mathrm{softmax}(\mathbf{O})$ to compute the probability distribution of the output categories.

This is entirely analogous to the regression problem we solved previously in Section 9.1, hence we omit details. Suffice it to say that we can pick feature-label pairs at random and learn the parameters of our network via automatic differentiation and stochastic gradient descent.

## 9.4.2 Recurrent Neural Networks with Hidden States

Matters are entirely different when we have hidden states. Let's look at the structure in some more detail.

Assume that we have a minibatch of inputs $\mathbf{X}_t \in \mathbb{R}^{n\times d}$ at time step $t$. In other words, for a minibatch of $n$ sequence examples, each row of $\mathbf{X}_t$ corresponds to one example at time step $t$ from the sequence. Next, denote by $\mathbf{H}_t \in \mathbb{R}^{n\times h}$ the hidden layer output of time step $t$. Unlike with MLP, here we save the hidden layer output $\mathbf{H}_{t-1}$ from the previous time step and introduce a new weight parameter $\mathbf{W}_{\mathrm{hh}} \in \mathbb{R}^{h\times h}$ to describe how to use the hidden layer output of the previous time step in the current time step. Specifically, the calculation of the hidden layer output of the current time step is determined by the input of the current time step together with the hidden layer output of the previous time step:

$$\mathbf{H}_t = \phi(\mathbf{X}_t\mathbf{W}_{\mathrm{xh}} + \mathbf{H}_{t-1}\mathbf{W}_{\mathrm{hh}} + \mathbf{b}_{\mathrm{h}}). \tag{9.4.5}$$

Compared with (9.4.3), (9.4.5) adds one more term $\mathbf{H}_{t-1}\mathbf{W}_{\mathrm{hh}}$ and thus instantiates (9.4.2). From the relationship between hidden layer outputs $\mathbf{H}_t$ and $\mathbf{H}_{t-1}$ of adjacent time steps, we know that these variables captured and retained the sequence's historical information up to their current time step, just like the state or memory of the neural network's current time step. Therefore, such a hidden layer output is called a *hidden state*. Since the hidden state uses the same definition of the previous time step in the current time step, the computation of (9.4.5) is *recurrent*. Hence, as we said, neural networks with hidden states based on recurrent computation are named *recurrent neural networks*. Layers that perform the computation of (9.4.5) in RNNs are called *recurrent layers*.

There are many different ways for constructing RNNs. Those with a hidden state defined by (9.4.5) are very common. For time step $t$, the output of the output layer is similar to the computation in the MLP:

$$\mathbf{O}_t = \mathbf{H}_t \mathbf{W}_{\textrm{hq}} + \mathbf{b}_\textrm{q}. \tag{9.4.6}$$

Parameters of the RNN include the weights $\mathbf{W}_{\textrm{xh}} \in \mathbb{R}^{d \times h}, \mathbf{W}_{\textrm{hh}} \in \mathbb{R}^{h \times h}$, and the bias $\mathbf{b}_\textrm{h} \in \mathbb{R}^{1 \times h}$ of the hidden layer, together with the weights $\mathbf{W}_{\textrm{hq}} \in \mathbb{R}^{h \times q}$ and the bias $\mathbf{b}_\textrm{q} \in \mathbb{R}^{1 \times q}$ of the output layer. It is worth mentioning that even at different time steps, RNNs always use these model parameters. Therefore, the parametrization cost of an RNN does not grow as the number of time steps increases.

Fig. 9.4.1 illustrates the computational logic of an RNN at three adjacent time steps. At any time step $t$, the computation of the hidden state can be treated as: (i) concatenating the input $\mathbf{X}_t$ at the current time step $t$ and the hidden state $\mathbf{H}_{t-1}$ at the previous time step $t-1$; (ii) feeding the concatenation result into a fully connected layer with the activation function $\phi$. The output of such a fully connected layer is the hidden state $\mathbf{H}_t$ of the current time step $t$. In this case, the model parameters are the concatenation of $\mathbf{W}_{\textrm{xh}}$ and $\mathbf{W}_{\textrm{hh}}$, and a bias of $\mathbf{b}_\textrm{h}$, all from (9.4.5). The hidden state of the current time step $t$, $\mathbf{H}_t$, will participate in computing the hidden state $\mathbf{H}_{t+1}$ of the next time step $t+1$. What is more, $\mathbf{H}_t$ will also be fed into the fully connected output layer to compute the output $\mathbf{O}_t$ of the current time step $t$.

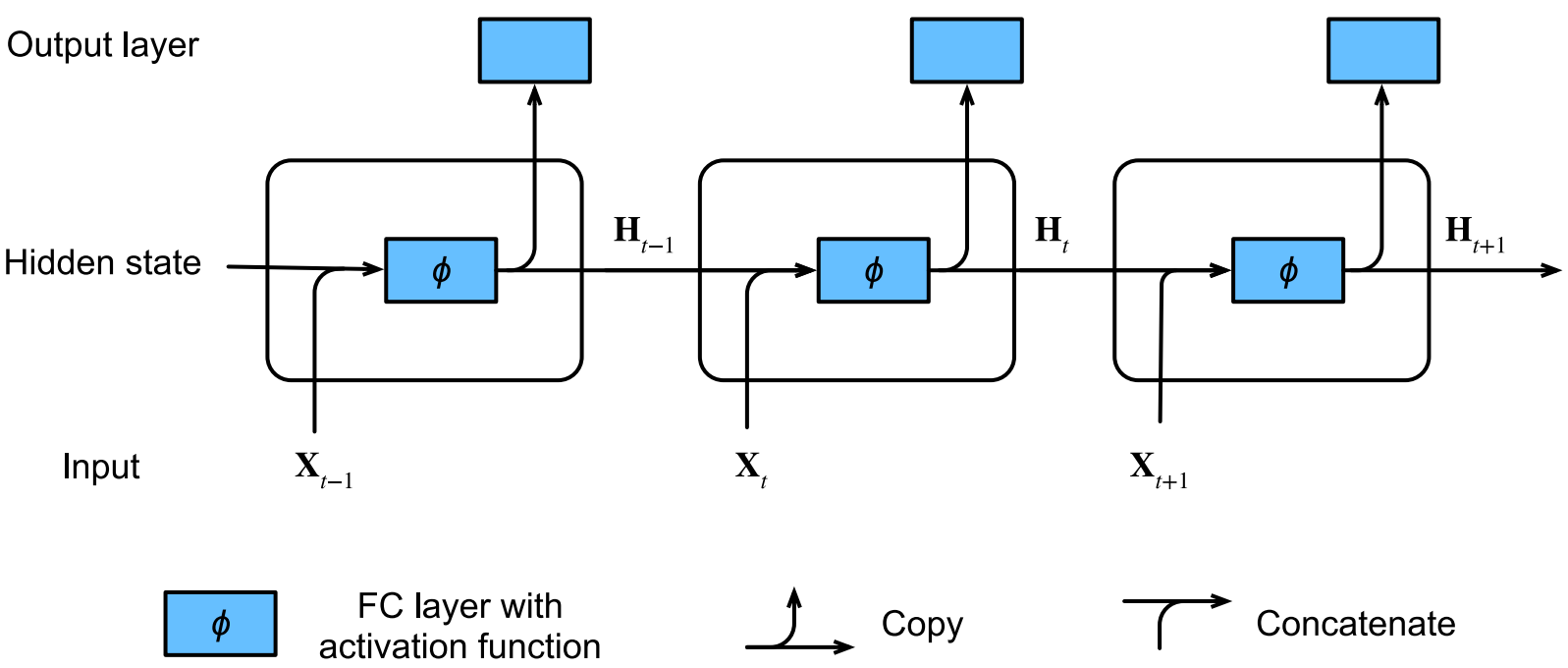

Fig. 9.4.1 An RNN with a hidden state.

We just mentioned that the calculation of $\mathbf{X}_t \mathbf{W}_{\textrm{xh}} + \mathbf{H}_{t-1} \mathbf{W}_{\textrm{hh}}$ for the hidden state is equivalent to matrix multiplication of the concatenation of $\mathbf{X}_t$ and $\mathbf{H}_{t-1}$ and the concatenation of $\mathbf{W}_{\textrm{xh}}$ and $\mathbf{W}_{\textrm{hh}}$. Though this can be proven mathematically, in the following we just use a simple code snippet as a demonstration. To begin with, we define matrices `X`, `W_xh`, `H`, and `W_hh`, whose shapes are (3, 1), (1, 4), (3, 4), and (4, 4), respectively. Multiplying `X` by `W_xh`, and `H` by `W_hh`, and then adding these two products, we obtain a matrix of shape (3, 4).

```
X, W_xh = torch.randn(3, 1), torch.randn(1, 4)
H, W_hh = torch.randn(3, 4), torch.randn(4, 4)
torch.matmul(X, W_xh) + torch.matmul(H, W_hh)
```

```
tensor([[-1.5596, -1.2137,  2.8331, -3.1458],
        [-2.2858,  7.0869,  0.7021, -3.2064],
        [ 0.4085, -2.1803, -1.2999,  1.1916]])
```

Now we concatenate the matrices `X` and `H` along columns (axis 1), and the matrices `W_xh` and `W_hh` along rows (axis 0). These two concatenations result in matrices of shape (3, 5) and of shape (5, 4), respectively. Multiplying these two concatenated matrices, we obtain the same output matrix of shape (3, 4) as above.

```
torch.matmul(torch.cat((X, H), 1), torch.cat((W_xh, W_hh), 0))
```

```
tensor([[-1.5596, -1.2137,  2.8331, -3.1459],
        [-2.2858,  7.0869,  0.7021, -3.2064],
        [ 0.4085, -2.1803, -1.2999,  1.1916]])
```

### 9.4.3 RNN-Based Character-Level Language Models

Recall that for language modeling in Section 9.3, we aim to predict the next token based on the current and past tokens; thus we shift the original sequence by one token as the targets (labels). Bengio *et al.* (2003) first proposed to use a neural network for language modeling. In the following we illustrate how RNNs can be used to build a language model. Let the minibatch size be one, and the sequence of the text be "machine". To simplify training in subsequent sections, we tokenize text into characters rather than words and consider a *character-level language model*. Fig. 9.4.2 demonstrates how to predict the next character based on the current and previous characters via an RNN for character-level language modeling.

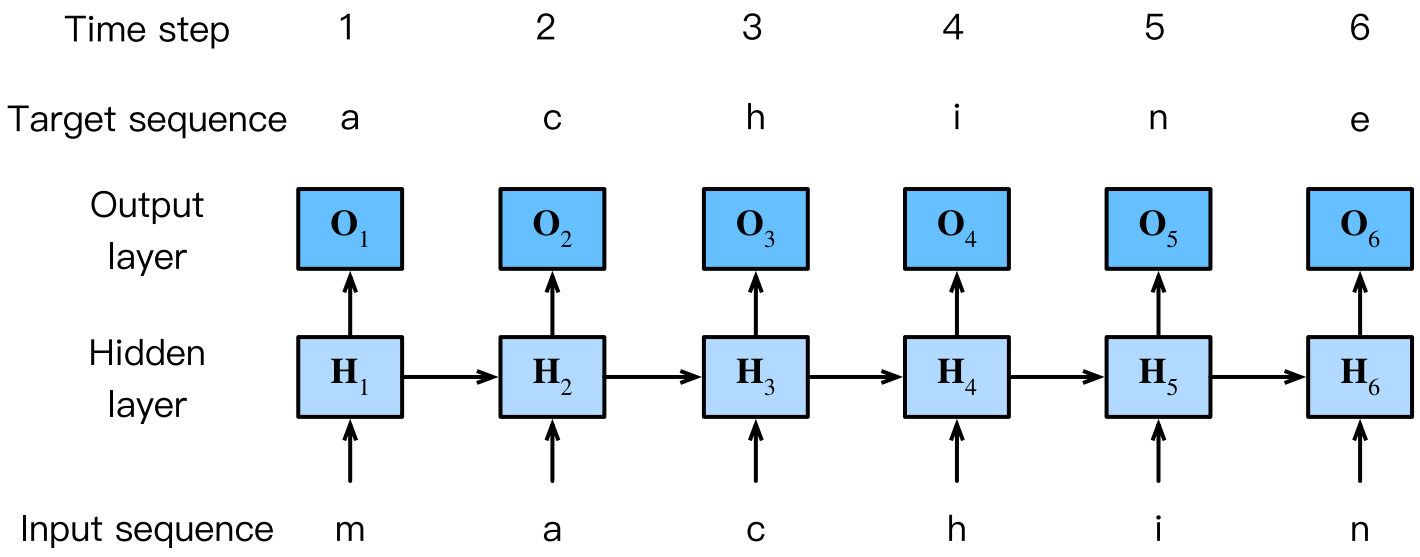

Fig. 9.4.2 A character-level language model based on the RNN. The input and target sequences are "machin" and "achine", respectively.

During the training process, we run a softmax operation on the output from the output layer for each time step, and then use the cross-entropy loss to compute the error between the model output and the target. Because of the recurrent computation of the hidden state in the hidden layer, the output, $\mathbf{O}_3$, of time step 3 in Fig. 9.4.2 is determined by the text sequence "m", "a", and "c". Since the next character of the sequence in the training data is "h", the loss of time step 3 will depend on the probability distribution of the next character generated based on the feature sequence "m", "a", "c" and the target "h" of this time step.

In practice, each token is represented by a $d$-dimensional vector, and we use a batch size $n > 1$. Therefore, the input $\mathbf{X}_t$ at time step $t$ will be an $n \times d$ matrix, which is identical to what we discussed in Section 9.4.2.

In the following sections, we will implement RNNs for character-level language models.

### 9.4.4 Summary

A neural network that uses recurrent computation for hidden states is called a recurrent neural network (RNN). The hidden state of an RNN can capture historical information of the sequence up to the current time step. With recurrent computation, the number of RNN model parameters does not grow as the number of time steps increases. As for applications, an RNN can be used to create character-level language models.

### 9.4.5 Exercises

1. If we use an RNN to predict the next character in a text sequence, what is the required dimension for any output?
2. Why can RNNs express the conditional probability of a token at some time step based on all the previous tokens in the text sequence?
3. What happens to the gradient if you backpropagate through a long sequence?
4. What are some of the problems associated with the language model described in this section?

Discussions[141].

## 9.5 Recurrent Neural Network Implementation from Scratch

We are now ready to implement an RNN from scratch. In particular, we will train this RNN to function as a character-level language model (see Section 9.4) and train it on a corpus consisting of the entire text of H. G. Wells' *The Time Machine*, following the data processing steps outlined in Section 9.2. We start by loading the dataset.

```
%matplotlib inline
import math
import torch
from torch import nn
from torch.nn import functional as F
from d2l import torch as d2l
```

## 9.5.1 RNN Model

We begin by defining a class to implement the RNN model (Section 9.4.2). Note that the number of hidden units `num_hiddens` is a tunable hyperparameter.

```
class RNNScratch(d2l.Module):  #@save
    """The RNN model implemented from scratch."""
    def __init__(self, num_inputs, num_hiddens, sigma=0.01):
        super().__init__()
        self.save_hyperparameters()
        self.W_xh = nn.Parameter(
            torch.randn(num_inputs, num_hiddens) * sigma)
        self.W_hh = nn.Parameter(
            torch.randn(num_hiddens, num_hiddens) * sigma)
        self.b_h = nn.Parameter(torch.zeros(num_hiddens))
```

The `forward` method below defines how to compute the output and hidden state at any time step, given the current input and the state of the model at the previous time step. Note that the RNN model loops through the outermost dimension of `inputs`, updating the hidden state one time step at a time. The model here uses a $\tanh$ activation function (Section 5.1.2).

```
@d2l.add_to_class(RNNScratch)  #@save
def forward(self, inputs, state=None):
    if state is None:
        # Initial state with shape: (batch_size, num_hiddens)
        state = torch.zeros((inputs.shape[1], self.num_hiddens),
                          device=inputs.device)
    else:
        state, = state
    outputs = []
    for X in inputs:  # Shape of inputs: (num_steps, batch_size, num_inputs)
        state = torch.tanh(torch.matmul(X, self.W_xh) +
                         torch.matmul(state, self.W_hh) + self.b_h)
        outputs.append(state)
    return outputs, state
```

We can feed a minibatch of input sequences into an RNN model as follows.

```
batch_size, num_inputs, num_hiddens, num_steps = 2, 16, 32, 100
rnn = RNNScratch(num_inputs, num_hiddens)
X = torch.ones((num_steps, batch_size, num_inputs))
outputs, state = rnn(X)
```

Let's check whether the RNN model produces results of the correct shapes to ensure that the dimensionality of the hidden state remains unchanged.

```
def check_len(a, n):  #@save
    """Check the length of a list."""
    assert len(a) == n, f'list\'s length {len(a)} != expected length {n}'

def check_shape(a, shape):  #@save
    """Check the shape of a tensor."""
    assert a.shape == shape, \
            f'tensor\'s shape {a.shape} != expected shape {shape}'

check_len(outputs, num_steps)
check_shape(outputs[0], (batch_size, num_hiddens))
check_shape(state, (batch_size, num_hiddens))
```

### 9.5.2 RNN-Based Language Model

The following `RNNLMScratch` class defines an RNN-based language model, where we pass in our RNN via the `rnn` argument of the `__init__` method. When training language models, the inputs and outputs are from the same vocabulary. Hence, they have the same dimension, which is equal to the vocabulary size. Note that we use perplexity to evaluate the model. As discussed in Section 9.3.2, this ensures that sequences of different length are comparable.

```
class RNNLMScratch(d2l.Classifier):  #@save
    """The RNN-based language model implemented from scratch."""
    def __init__(self, rnn, vocab_size, lr=0.01):
        super().__init__()
        self.save_hyperparameters()
        self.init_params()

    def init_params(self):
        self.W_hq = nn.Parameter(
            torch.randn(
                self.rnn.num_hiddens, self.vocab_size) * self.rnn.sigma)
        self.b_q = nn.Parameter(torch.zeros(self.vocab_size))

    def training_step(self, batch):
        l = self.loss(self(*batch[:-1]), batch[-1])
        self.plot('ppl', torch.exp(l), train=True)
        return l

    def validation_step(self, batch):
        l = self.loss(self(*batch[:-1]), batch[-1])
        self.plot('ppl', torch.exp(l), train=False)
```

### One-Hot Encoding

Recall that each token is represented by a numerical index indicating the position in the vocabulary of the corresponding word/character/word piece. You might be tempted to build a neural network with a single input node (at each time step), where the index could be fed in as a scalar value. This works when we are dealing with numerical inputs like price or temperature, where any two values sufficiently close together should be treated similarly. But this does not quite make sense. The $45^{\mathrm{th}}$ and $46^{\mathrm{th}}$ words in our vocabulary happen to be “their” and “said”, whose meanings are not remotely similar.

When dealing with such categorical data, the most common strategy is to represent each item by a *one-hot encoding* (recall from Section 4.1.1). A one-hot encoding is a vector whose length is given by the size of the vocabulary $N$, where all entries are set to 0, except for the entry corresponding to our token, which is set to 1. For example, if the vocabulary had five elements, then the one-hot vectors corresponding to indices 0 and 2 would be the following.

```
F.one_hot(torch.tensor([0, 2]), 5)
```

```
tensor([[1, 0, 0, 0, 0],
        [0, 0, 1, 0, 0]])
```

The minibatches that we sample at each iteration will take the shape (batch size, number of time steps). Once representing each input as a one-hot vector, we can think of each minibatch as a three-dimensional tensor, where the length along the third axis is given by the vocabulary size (`len(vocab)`). We often transpose the input so that we will obtain an output of shape (number of time steps, batch size, vocabulary size). This will allow us to loop more conveniently through the outermost dimension for updating hidden states of a minibatch, time step by time step (e.g., in the above `forward` method).

```
@d2l.add_to_class(RNNLMScratch)  #@save
def one_hot(self, X):
    # Output shape: (num_steps, batch_size, vocab_size)
    return F.one_hot(X.T, self.vocab_size).type(torch.float32)
```

### Transforming RNN Outputs

The language model uses a fully connected output layer to transform RNN outputs into token predictions at each time step.

```
@d2l.add_to_class(RNNLMScratch)  #@save
def output_layer(self, rnn_outputs):
    outputs = [torch.matmul(H, self.W_hq) + self.b_q for H in rnn_outputs]
    return torch.stack(outputs, 1)

@d2l.add_to_class(RNNLMScratch)  #@save
```

```
def forward(self, X, state=None):
    embs = self.one_hot(X)
    rnn_outputs, _ = self.rnn(embs, state)
    return self.output_layer(rnn_outputs)
```

Let's check whether the forward computation produces outputs with the correct shape.

```
model = RNNLMScratch(rnn, num_inputs)
outputs = model(torch.ones((batch_size, num_steps), dtype=torch.int64))
check_shape(outputs, (batch_size, num_steps, num_inputs))
```

### 9.5.3 Gradient Clipping

While you are already used to thinking of neural networks as "deep" in the sense that many layers separate the input and output even within a single time step, the length of the sequence introduces a new notion of depth. In addition to the passing through the network in the input-to-output direction, inputs at the first time step must pass through a chain of $T$ layers along the time steps in order to influence the output of the model at the final time step. Taking the backwards view, in each iteration, we backpropagate gradients through time, resulting in a chain of matrix-products of length $O(T)$. As mentioned in Section 5.4, this can result in numerical instability, causing the gradients either to explode or vanish, depending on the properties of the weight matrices.

Dealing with vanishing and exploding gradients is a fundamental problem when designing RNNs and has inspired some of the biggest advances in modern neural network architectures. In the next chapter, we will talk about specialized architectures that were designed in hopes of mitigating the vanishing gradient problem. However, even modern RNNs often suffer from exploding gradients. One inelegant but ubiquitous solution is to simply clip the gradients forcing the resulting "clipped" gradients to take smaller values.

Generally speaking, when optimizing some objective by gradient descent, we iteratively update the parameter of interest, say a vector $\mathbf{x}$, but pushing it in the direction of the negative gradient $\mathbf{g}$ (in stochastic gradient descent, we calculate this gradient on a randomly sampled minibatch). For example, with learning rate $\eta > 0$, each update takes the form $\mathbf{x} \leftarrow \mathbf{x} - \eta\mathbf{g}$. Let's further assume that the objective function $f$ is sufficiently smooth. Formally, we say that the objective is *Lipschitz continuous* with constant $L$, meaning that for any $\mathbf{x}$ and $\mathbf{y}$, we have

$$|f(\mathbf{x}) - f(\mathbf{y})| \leq L\|\mathbf{x} - \mathbf{y}\|. \tag{9.5.1}$$

As you can see, when we update the parameter vector by subtracting $\eta\mathbf{g}$, the change in the value of the objective depends on the learning rate, the norm of the gradient and $L$ as follows:

$$|f(\mathbf{x}) - f(\mathbf{x} - \eta\mathbf{g})| \leq L\eta\|\mathbf{g}\|. \tag{9.5.2}$$

In other words, the objective cannot change by more than $L\eta\|\mathbf{g}\|$. Having a small value for this upper bound might be viewed as good or bad. On the downside, we are limiting the speed at which we can reduce the value of the objective. On the bright side, this limits by just how much we can go wrong in any one gradient step.

When we say that gradients explode, we mean that $\|\mathbf{g}\|$ becomes excessively large. In this worst case, we might do so much damage in a single gradient step that we could undo all of the progress made over the course of thousands of training iterations. When gradients can be so large, neural network training often diverges, failing to reduce the value of the objective. At other times, training eventually converges but is unstable owing to massive spikes in the loss.

One way to limit the size of $L\eta\|\mathbf{g}\|$ is to shrink the learning rate $\eta$ to tiny values. This has the advantage that we do not bias the updates. But what if we only *rarely* get large gradients? This drastic move slows down our progress at all steps, just to deal with the rare exploding gradient events. A popular alternative is to adopt a *gradient clipping* heuristic projecting the gradients $\mathbf{g}$ onto a ball of some given radius $\theta$ as follows:

$$\mathbf{g} \leftarrow \min\left(1, \frac{\theta}{\|\mathbf{g}\|}\right) \mathbf{g}. \tag{9.5.3}$$

This ensures that the gradient norm never exceeds $\theta$ and that the updated gradient is entirely aligned with the original direction of $\mathbf{g}$. It also has the desirable side-effect of limiting the influence any given minibatch (and within it any given sample) can exert on the parameter vector. This bestows a certain degree of robustness to the model. To be clear, it is a hack. Gradient clipping means that we are not always following the true gradient and it is hard to reason analytically about the possible side effects. However, it is a very useful hack, and is widely adopted in RNN implementations in most deep learning frameworks.

Below we define a method to clip gradients, which is invoked by the `fit_epoch` method of the `d2l.Trainer` class (see Section 3.4). Note that when computing the gradient norm, we are concatenating all model parameters, treating them as a single giant parameter vector.

```
@d2l.add_to_class(d2l.Trainer)  #@save
def clip_gradients(self, grad_clip_val, model):
    params = [p for p in model.parameters() if p.requires_grad]
    norm = torch.sqrt(sum(torch.sum((p.grad ** 2)) for p in params))
    if norm > grad_clip_val:
        for param in params:
            param.grad[:] *= grad_clip_val / norm
```

## 9.5.4 Training

Using *The Time Machine* dataset (`data`), we train a character-level language model (`model`) based on the RNN (`rnn`) implemented from scratch. Note that we first calculate the gradients, then clip them, and finally update the model parameters using the clipped gradients.

```
data = d2l.TimeMachine(batch_size=1024, num_steps=32)
rnn = RNNScratch(num_inputs=len(data.vocab), num_hiddens=32)
model = RNNLMScratch(rnn, vocab_size=len(data.vocab), lr=1)
trainer = d2l.Trainer(max_epochs=100, gradient_clip_val=1, num_gpus=1)
trainer.fit(model, data)
```

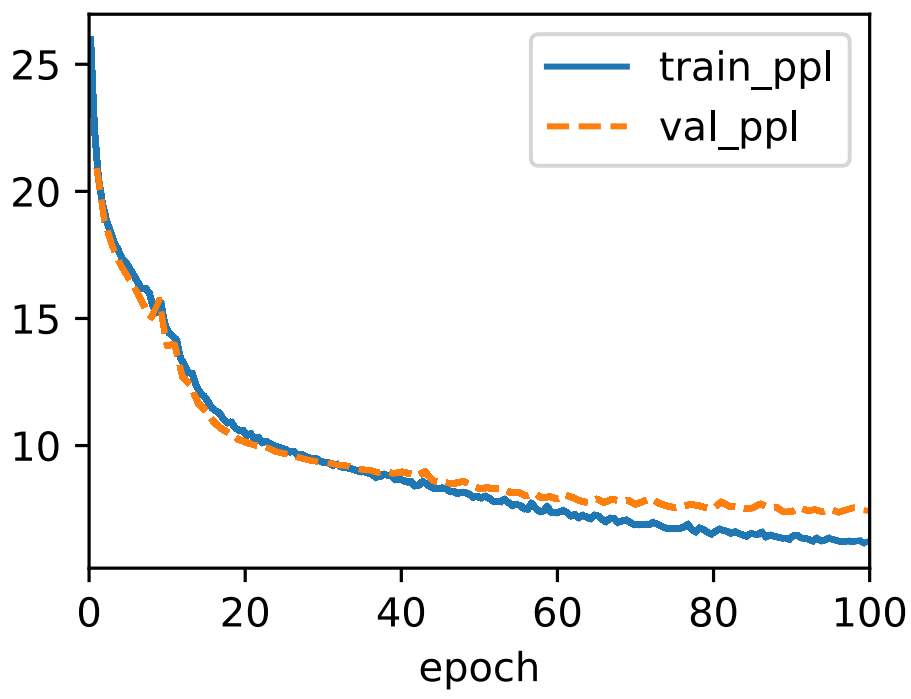

## 9.5.5 Decoding

Once a language model has been learned, we can use it not only to predict the next token but to continue predicting each subsequent one, treating the previously predicted token as though it were the next in the input. Sometimes we will just want to generate text as though we were starting at the beginning of a document. However, it is often useful to condition the language model on a user-supplied prefix. For example, if we were developing an autocomplete feature for a search engine or to assist users in writing emails, we would want to feed in what they had written so far (the prefix), and then generate a likely continuation.

The following `predict` method generates a continuation, one character at a time, after ingesting a user-provided `prefix`. When looping through the characters in `prefix`, we keep passing the hidden state to the next time step but do not generate any output. This is called the *warm-up* period. After ingesting the prefix, we are now ready to begin emitting the subsequent characters, each of which will be fed back into the model as the input at the next time step.

```
@d2l.add_to_class(RNNLMScratch)  #@save
def predict(self, prefix, num_preds, vocab, device=None):
    state, outputs = None, [vocab[prefix[0]]]
    for i in range(len(prefix) + num_preds - 1):
        X = torch.tensor([[outputs[-1]]], device=device)
        embs = self.one_hot(X)
        rnn_outputs, state = self.rnn(embs, state)
        if i < len(prefix) - 1:  # Warm-up period
            outputs.append(vocab[prefix[i + 1]])
        else:  # Predict num_preds steps
            Y = self.output_layer(rnn_outputs)
```

```
            outputs.append(int(Y.argmax(axis=2).reshape(1)))
    return ''.join([vocab.idx_to_token[i] for i in outputs])
```

In the following, we specify the prefix and have it generate 20 additional characters.

```
model.predict('it has', 20, data.vocab, d2l.try_gpu())
```

```
'it has is a mere and the t'
```

While implementing the above RNN model from scratch is instructive, it is not convenient. In the next section, we will see how to leverage deep learning frameworks to whip up RNNs using standard architectures, and to reap performance gains by relying on highly optimized library functions.

## 9.5.6 Summary

We can train RNN-based language models to generate text following the user-provided text prefix. A simple RNN language model consists of input encoding, RNN modeling, and output generation. During training, gradient clipping can mitigate the problem of exploding gradients but does not address the problem of vanishing gradients. In the experiment, we implemented a simple RNN language model and trained it with gradient clipping on sequences of text, tokenized at the character level. By conditioning on a prefix, we can use a language model to generate likely continuations, which proves useful in many applications, e.g., autocomplete features.

## 9.5.7 Exercises

1. Does the implemented language model predict the next token based on all the past tokens up to the very first token in *The Time Machine*?
2. Which hyperparameter controls the length of history used for prediction?
3. Show that one-hot encoding is equivalent to picking a different embedding for each object.
4. Adjust the hyperparameters (e.g., number of epochs, number of hidden units, number of time steps in a minibatch, and learning rate) to improve the perplexity. How low can you go while sticking with this simple architecture?
5. Replace one-hot encoding with learnable embeddings. Does this lead to better performance?
6. Conduct an experiment to determine how well this language model trained on *The Time Machine* works on other books by H. G. Wells, e.g., *The War of the Worlds*.

7. Conduct another experiment to evaluate the perplexity of this model on books written by other authors.
8. Modify the prediction method so as to use sampling rather than picking the most likely next character.
   - What happens?
   - Bias the model towards more likely outputs, e.g., by sampling from $q(x_t \mid x_{t-1}, \ldots, x_1) \propto P(x_t \mid x_{t-1}, \ldots, x_1)^\alpha$ for $\alpha > 1$.
9. Run the code in this section without clipping the gradient. What happens?
10. Replace the activation function used in this section with ReLU and repeat the experiments in this section. Do we still need gradient clipping? Why?

Discussions[142].

# 9.6 Concise Implementation of Recurrent Neural Networks

Like most of our from-scratch implementations, Section 9.5 was designed to provide insight into how each component works. But when you are using RNNs every day or writing production code, you will want to rely more on libraries that cut down on both implementation time (by supplying library code for common models and functions) and computation time (by optimizing the heck out of these library implementations). This section will show you how to implement the same language model more efficiently using the high-level API provided by your deep learning framework. We begin, as before, by loading *The Time Machine* dataset.

```
import torch
from torch import nn
from torch.nn import functional as F
from d2l import torch as d2l
```

## 9.6.1 Defining the Model

We define the following class using the RNN implemented by high-level APIs.

```
class RNN(d2l.Module):  #@save
    """The RNN model implemented with high-level APIs."""
    def __init__(self, num_inputs, num_hiddens):
        super().__init__()
```

```
        self.save_hyperparameters()
        self.rnn = nn.RNN(num_inputs, num_hiddens)

    def forward(self, inputs, H=None):
        return self.rnn(inputs, H)
```

Inheriting from the `RNNLMScratch` class in Section 9.5, the following `RNNLM` class defines a complete RNN-based language model. Note that we need to create a separate fully connected output layer.

```
class RNNLM(d2l.RNNLMScratch):  #@save
    """The RNN-based language model implemented with high-level APIs."""
    def init_params(self):
        self.linear = nn.LazyLinear(self.vocab_size)

    def output_layer(self, hiddens):
        return self.linear(hiddens).swapaxes(0, 1)
```

## 9.6.2 Training and Predicting

Before training the model, let's make a prediction with a model initialized with random weights. Given that we have not trained the network, it will generate nonsensical predictions.

```
data = d2l.TimeMachine(batch_size=1024, num_steps=32)
rnn = RNN(num_inputs=len(data.vocab), num_hiddens=32)
model = RNNLM(rnn, vocab_size=len(data.vocab), lr=1)
model.predict('it has', 20, data.vocab)
```

```
'it hassuussuussuussuussuus'
```

Next, we train our model, leveraging the high-level API.

```
trainer = d2l.Trainer(max_epochs=100, gradient_clip_val=1, num_gpus=1)
trainer.fit(model, data)
```

Compared with Section 9.5, this model achieves comparable perplexity, but runs faster due to the optimized implementations. As before, we can generate predicted tokens following the specified prefix string.

```
model.predict('it has', 20, data.vocab, d2l.try_gpu())
```

```
'it has a mather a mather a'
```

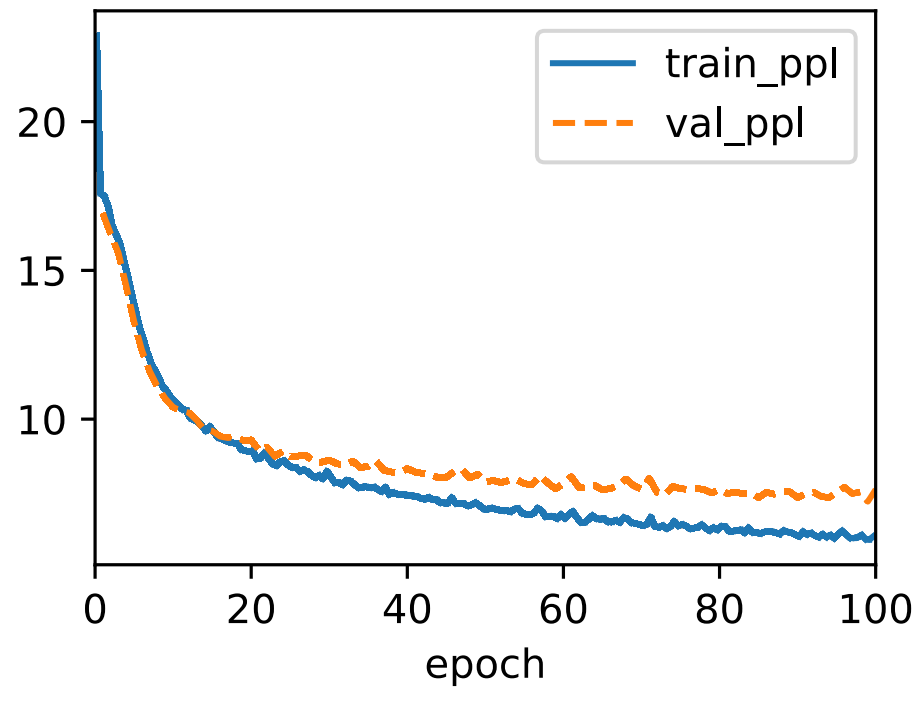

### 9.6.3 Summary

High-level APIs in deep learning frameworks provide implementations of standard RNNs. These libraries help you to avoid wasting time reimplementing standard models. Moreover, framework implementations are often highly optimized, leading to significant (computational) performance gains when compared with implementations from scratch.

### 9.6.4 Exercises

1. Can you make the RNN model overfit using the high-level APIs?
2. Implement the autoregressive model of Section 9.1 using an RNN.

Discussions[143].

## 9.7 Backpropagation Through Time

If you completed the exercises in Section 9.5, you would have seen that gradient clipping is vital for preventing the occasional massive gradients from destabilizing training. We hinted that the exploding gradients stem from backpropagating across long sequences. Before introducing a slew of modern RNN architectures, let's take a closer look at how *backpropagation* works in sequence models in mathematical detail. Hopefully, this discussion will bring some precision to the notion of *vanishing* and *exploding* gradients. If you recall our discussion of forward and backward propagation through computational graphs when we introduced MLPs in Section 5.3, then forward propagation in RNNs should be relatively straightforward. Applying backpropagation in RNNs is called *backpropagation through time* (Werbos, 1990). This procedure requires us to expand (or unroll) the computational graph of an RNN one time step at a time. The unrolled RNN is essentially a feedforward neural network with the special property that the same parameters are repeated throughout the unrolled network, appearing at each time step. Then, just as in any feedforward neural network, we can apply the chain rule, backpropagating gradients through the unrolled net. The gradient with respect to each parameter must be summed across all places that the parameter occurs in the unrolled net. Handling such weight tying should be familiar from our chapters on convolutional neural networks.

Complications arise because sequences can be rather long. It is not unusual to work with text sequences consisting of over a thousand tokens. Note that this poses problems both from a computational (too much memory) and optimization (numerical instability) standpoint. Input from the first step passes through over 1000 matrix products before arriving at the output, and another 1000 matrix products are required to compute the gradient. We now analyze what can go wrong and how to address it in practice.

## 9.7.1 Analysis of Gradients in RNNs

We start with a simplified model of how an RNN works. This model ignores details about the specifics of the hidden state and how it is updated. The mathematical notation here does not explicitly distinguish scalars, vectors, and matrices. We are just trying to develop some intuition. In this simplified model, we denote $h_t$ as the hidden state, $x_t$ as input, and $o_t$ as output at time step $t$. Recall our discussions in Section 9.4.2 that the input and the hidden state can be concatenated before being multiplied by one weight variable in the hidden layer. Thus, we use $w_\textrm{h}$ and $w_\textrm{o}$ to indicate the weights of the hidden layer and the output layer, respectively. As a result, the hidden states and outputs at each time step are

$$\begin{aligned}h_t &= f(x_t,h_{t-1},w_\textrm{h}),\\o_t &= g(h_t, w_\textrm{o}),\end{aligned} \tag{9.7.1}$$

where $f$ and $g$ are transformations of the hidden layer and the output layer, respectively. Hence, we have a chain of values $\{\ldots, (x_{t-1}, h_{t-1}, o_{t-1}), (x_{t}, h_{t}, o_t), \ldots\}$ that depend on each other via recurrent computation. The forward propagation is fairly straightforward. All we need is to loop through the $(x_t, h_t, o_t)$ triples one time step at a time. The discrepancy between output $o_t$ and the desired target $y_t$ is then evaluated by an objective function across all the $T$ time steps as

$$L(x_1, \ldots, x_T, y_1, \ldots, y_T, w_\textrm{h}, w_\textrm{o}) = \frac{1}{T}\sum_{t=1}^T l(y_t, o_t). \tag{9.7.2}$$

For backpropagation, matters are a bit trickier, especially when we compute the gradients with regard to the parameters $w_\textrm{h}$ of the objective function $L$. To be specific, by the chain rule,

$$\begin{aligned}\frac{\partial L}{\partial w_\textrm{h}} & = \frac{1}{T}\sum_{t=1}^T \frac{\partial l(y_t, o_t)}{\partial w_\textrm{h}} \\& = \frac{1}{T}\sum_{t=1}^T \frac{\partial l(y_t, o_t)}{\partial o_t} \frac{\partial g(h_t, w_\textrm{o})}{\partial h_t} \frac{\partial h_t}{\partial w_\textrm{h}}.\end{aligned} \tag{9.7.3}$$

The first and the second factors of the product in (9.7.3) are easy to compute. The third factor $\partial h_t/\partial w_\textrm{h}$ is where things get tricky, since we need to recurrently compute the effect of the parameter $w_\textrm{h}$ on $h_t$. According to the recurrent computation in (9.7.1), $h_t$ depends on both $h_{t-1}$ and $w_\textrm{h}$, where computation of $h_{t-1}$ also depends on $w_\textrm{h}$. Thus, evaluating the total derivate of $h_t$ with respect to $w_\textrm{h}$ using the chain rule yields

$$\frac{\partial h_t}{\partial w_\textrm{h}}= \frac{\partial f(x_{t},h_{t-1},w_\textrm{h})}{\partial w_\textrm{h}} +\frac{\partial f(x_{t},h_{t-1},w_\textrm{h})}{\partial h_{t-1}} \frac{\partial h_{t-1}}{\partial w_\textrm{h}}. \tag{9.7.4}$$

To derive the above gradient, assume that we have three sequences $\{a_{t}\},\{b_{t}\},\{c_{t}\}$ satisfying $a_{0}=0$ and $a_{t}=b_{t}+c_{t}a_{t-1}$ for $t=1, 2,\ldots$. Then for $t\geq 1$, it is easy to show

$$a_{t}=b_{t}+\sum_{i=1}^{t-1}\left(\prod_{j=i+1}^{t}c_{j}\right)b_{i}. \tag{9.7.5}$$

By substituting $a_t$, $b_t$, and $c_t$ according to

$$\begin{aligned}a_t &= \frac{\partial h_t}{\partial w_\textrm{h}},\\
b_t &= \frac{\partial f(x_{t},h_{t-1},w_\textrm{h})}{\partial w_\textrm{h}}, \\
c_t &= \frac{\partial f(x_{t},h_{t-1},w_\textrm{h})}{\partial h_{t-1}},\end{aligned} \tag{9.7.6}$$

the gradient computation in (9.7.4) satisfies $a_{t}=b_{t}+c_{t}a_{t-1}$. Thus, per (9.7.5), we can remove the recurrent computation in (9.7.4) with

$$\frac{\partial h_t}{\partial w_\textrm{h}}=\frac{\partial f(x_{t},h_{t-1},w_\textrm{h})}{\partial w_\textrm{h}}+\sum_{i=1}^{t-1}\left(\prod_{j=i+1}^{t} \frac{\partial f(x_{j},h_{j-1},w_\textrm{h})}{\partial h_{j-1}} \right) \frac{\partial f(x_{i},h_{i-1},w_\textrm{h})}{\partial w_\textrm{h}}. \tag{9.7.7}$$

While we can use the chain rule to compute $\partial h_t/\partial w_\textrm{h}$ recursively, this chain can get very long whenever $t$ is large. Let's discuss a number of strategies for dealing with this problem.

## Full Computation

One idea might be to compute the full sum in (9.7.7). However, this is very slow and gradients can blow up, since subtle changes in the initial conditions can potentially affect the outcome a lot. That is, we could see things similar to the butterfly effect, where minimal changes in the initial conditions lead to disproportionate changes in the outcome. This is generally undesirable. After all, we are looking for robust estimators that generalize well. Hence this strategy is almost never used in practice.

## Truncating Time Steps

Alternatively, we can truncate the sum in (9.7.7) after $\tau$ steps. This is what we have been discussing so far. This leads to an *approximation* of the true gradient, simply by terminating the sum at $\partial h_{t-\tau}/\partial w_\textrm{h}$. In practice this works quite well. It is what is commonly referred to as truncated backpropgation through time (Jaeger, 2002). One of the consequences of this is that the model focuses primarily on short-term influence rather than long-term consequences. This is actually *desirable*, since it biases the estimate towards simpler and more stable models.

### Randomized Truncation

Last, we can replace $\partial h_t/\partial w_\mathrm{h}$ by a random variable which is correct in expectation but truncates the sequence. This is achieved by using a sequence of $\xi_t$ with predefined $0 \leq \pi_t \leq 1$, where $P(\xi_t = 0) = 1 - \pi_t$ and $P(\xi_t = \pi_t^{-1}) = \pi_t$, thus $E[\xi_t] = 1$. We use this to replace the gradient $\partial h_t/\partial w_\mathrm{h}$ in (9.7.4) with

$$z_t = \frac{\partial f(x_t, h_{t-1}, w_\mathrm{h})}{\partial w_\mathrm{h}} + \xi_t \frac{\partial f(x_t, h_{t-1}, w_\mathrm{h})}{\partial h_{t-1}} \frac{\partial h_{t-1}}{\partial w_\mathrm{h}}. \tag{9.7.8}$$

It follows from the definition of $\xi_t$ that $E[z_t] = \partial h_t/\partial w_\mathrm{h}$. Whenever $\xi_t = 0$ the recurrent computation terminates at that time step $t$. This leads to a weighted sum of sequences of varying lengths, where long sequences are rare but appropriately overweighted. This idea was proposed by Tallec and Ollivier (2017).

### Comparing Strategies

`the time machine by h g wells`

Fig. 9.7.1 Comparing strategies for computing gradients in RNNs. From top to bottom: randomized truncation, regular truncation, and full computation.

Fig. 9.7.1 illustrates the three strategies when analyzing the first few characters of *The Time Machine* using backpropagation through time for RNNs:

- The first row is the randomized truncation that partitions the text into segments of varying lengths.
- The second row is the regular truncation that breaks the text into subsequences of the same length. This is what we have been doing in RNN experiments.
- The third row is the full backpropagation through time that leads to a computationally infeasible expression.

Unfortunately, while appealing in theory, randomized truncation does not work much better than regular truncation, most likely due to a number of factors. First, the effect of an observation after a number of backpropagation steps into the past is quite sufficient to capture dependencies in practice. Second, the increased variance counteracts the fact that the gradient is more accurate with more steps. Third, we actually *want* models that have only a short range of interactions. Hence, regularly truncated backpropagation through time has a slight regularizing effect that can be desirable.

### 9.7.2 Backpropagation Through Time in Detail

After discussing the general principle, let's discuss backpropagation through time in detail. In contrast to the analysis in Section 9.7.1, in the following we will show how to compute the gradients of the objective function with respect to all the decomposed model parameters. To keep things simple, we consider an RNN without bias parameters, whose activation function in the hidden layer uses the identity mapping ($\phi(x) = x$). For time step $t$, let the single example input and the target be $\mathbf{x}_t \in \mathbb{R}^d$ and $y_t$, respectively. The hidden state $\mathbf{h}_t \in \mathbb{R}^h$ and the output $\mathbf{o}_t \in \mathbb{R}^q$ are computed as

$$\begin{aligned}\mathbf{h}_t &= \mathbf{W}_{\text{hx}} \mathbf{x}_t + \mathbf{W}_{\text{hh}} \mathbf{h}_{t-1},\\ \mathbf{o}_t &= \mathbf{W}_{\text{qh}} \mathbf{h}_{t},\end{aligned} \tag{9.7.9}$$

where $\mathbf{W}_{\text{hx}} \in \mathbb{R}^{h \times d}$, $\mathbf{W}_{\text{hh}} \in \mathbb{R}^{h \times h}$, and $\mathbf{W}_{\text{qh}} \in \mathbb{R}^{q \times h}$ are the weight parameters. Denote by $l(\mathbf{o}_t, y_t)$ the loss at time step $t$. Our objective function, the loss over $T$ time steps from the beginning of the sequence is thus

$$L = \frac{1}{T} \sum_{t=1}^T l(\mathbf{o}_t, y_t). \tag{9.7.10}$$

In order to visualize the dependencies among model variables and parameters during computation of the RNN, we can draw a computational graph for the model, as shown in Fig. 9.7.2. For example, the computation of the hidden states of time step 3, $\mathbf{h}_3$, depends on the model parameters $\mathbf{W}_{\text{hx}}$ and $\mathbf{W}_{\text{hh}}$, the hidden state of the previous time step $\mathbf{h}_2$, and the input of the current time step $\mathbf{x}_3$.

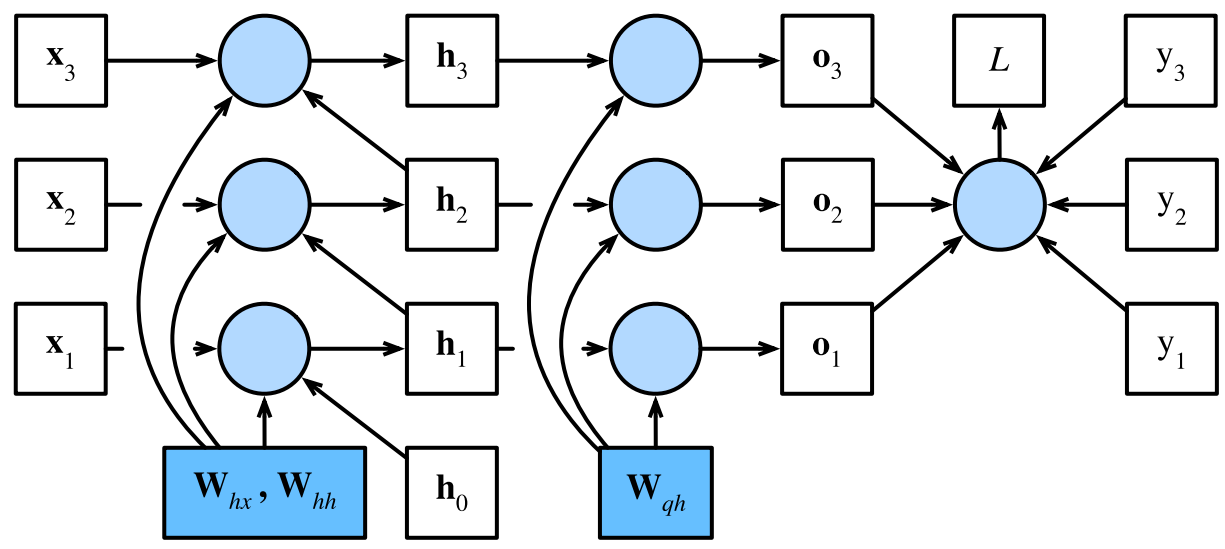

Fig. 9.7.2 Computational graph showing dependencies for an RNN model with three time steps. Boxes represent variables (not shaded) or parameters (shaded) and circles represent operators.

As just mentioned, the model parameters in Fig. 9.7.2 are $\mathbf{W}_{\text{hx}}$, $\mathbf{W}_{\text{hh}}$, and $\mathbf{W}_{\text{qh}}$. Generally, training this model requires gradient computation with respect to these parameters $\partial L/\partial \mathbf{W}_{\text{hx}}$, $\partial L/\partial \mathbf{W}_{\text{hh}}$, and $\partial L/\partial \mathbf{W}_{\text{qh}}$. According to the dependencies in Fig. 9.7.2, we can traverse in the opposite direction of the arrows to calculate and store the gradients in turn. To flexibly express the multiplication of matrices, vectors, and scalars of different shapes in the chain rule, we continue to use the prod operator as described in Section 5.3.

First of all, differentiating the objective function with respect to the model output at any time step $t$ is fairly straightforward:

$$\frac{\partial L}{\partial \mathbf{o}_t} = \frac{\partial l(\mathbf{o}_t, y_t)}{T \cdot \partial \mathbf{o}_t} \in \mathbb{R}^q. \tag{9.7.11}$$

Now we can calculate the gradient of the objective with respect to the parameter $\mathbf{W}_{\textrm{qh}}$ in the output layer: $\partial L/\partial \mathbf{W}_{\textrm{qh}} \in \mathbb{R}^{q \times h}$. Based on Fig. 9.7.2, the objective $L$ depends on $\mathbf{W}_{\textrm{qh}}$ via $\mathbf{o}_1, \ldots, \mathbf{o}_T$. Using the chain rule yields

$$\frac{\partial L}{\partial \mathbf{W}_{\textrm{qh}}} = \sum_{t=1}^T \textrm{prod}\left(\frac{\partial L}{\partial \mathbf{o}_t}, \frac{\partial \mathbf{o}_t}{\partial \mathbf{W}_{\textrm{qh}}}\right) = \sum_{t=1}^T \frac{\partial L}{\partial \mathbf{o}_t} \mathbf{h}_t^\top, \tag{9.7.12}$$

where $\partial L/\partial \mathbf{o}_t$ is given by (9.7.11).

Next, as shown in Fig. 9.7.2, at the final time step $T$, the objective function $L$ depends on the hidden state $\mathbf{h}_T$ only via $\mathbf{o}_T$. Therefore, we can easily find the gradient $\partial L/\partial \mathbf{h}_T \in \mathbb{R}^h$ using the chain rule:

$$\frac{\partial L}{\partial \mathbf{h}_T} = \textrm{prod}\left(\frac{\partial L}{\partial \mathbf{o}_T}, \frac{\partial \mathbf{o}_T}{\partial \mathbf{h}_T}\right) = \mathbf{W}_{\textrm{qh}}^\top \frac{\partial L}{\partial \mathbf{o}_T}. \tag{9.7.13}$$

It gets trickier for any time step $t < T$, where the objective function $L$ depends on $\mathbf{h}_t$ via $\mathbf{h}_{t+1}$ and $\mathbf{o}_t$. According to the chain rule, the gradient of the hidden state $\partial L/\partial \mathbf{h}_t \in \mathbb{R}^h$ at any time step $t < T$ can be recurrently computed as:

$$\frac{\partial L}{\partial \mathbf{h}_t} = \textrm{prod}\left(\frac{\partial L}{\partial \mathbf{h}_{t+1}}, \frac{\partial \mathbf{h}_{t+1}}{\partial \mathbf{h}_t}\right) + \textrm{prod}\left(\frac{\partial L}{\partial \mathbf{o}_t}, \frac{\partial \mathbf{o}_t}{\partial \mathbf{h}_t}\right) = \mathbf{W}_{\textrm{hh}}^\top \frac{\partial L}{\partial \mathbf{h}_{t+1}} + \mathbf{W}_{\textrm{qh}}^\top \frac{\partial L}{\partial \mathbf{o}_t}. \tag{9.7.14}$$

For analysis, expanding the recurrent computation for any time step $1 \leq t \leq T$ gives

$$\frac{\partial L}{\partial \mathbf{h}_t} = \sum_{i=t}^T {\left(\mathbf{W}_{\textrm{hh}}^\top\right)}^{T-i} \mathbf{W}_{\textrm{qh}}^\top \frac{\partial L}{\partial \mathbf{o}_{T+t-i}}. \tag{9.7.15}$$

We can see from (9.7.15) that this simple linear example already exhibits some key problems of long sequence models: it involves potentially very large powers of $\mathbf{W}_{\textrm{hh}}^\top$. In it, eigenvalues smaller than 1 vanish and eigenvalues larger than 1 diverge. This is numerically unstable, which manifests itself in the form of vanishing and exploding gradients. One way to address this is to truncate the time steps at a computationally convenient size as discussed in Section 9.7.1. In practice, this truncation can also be effected by detaching the gradient after a given number of time steps. Later on, we will see how more sophisticated sequence models such as long short-term memory can alleviate this further.

Finally, Fig. 9.7.2 shows that the objective function $L$ depends on model parameters $\mathbf{W}_{\textrm{hx}}$ and $\mathbf{W}_{\textrm{hh}}$ in the hidden layer via hidden states $\mathbf{h}_1, \ldots, \mathbf{h}_T$. To compute gradients with respect to such parameters $\partial L / \partial \mathbf{W}_{\textrm{hx}} \in \mathbb{R}^{h \times d}$ and $\partial L / \partial \mathbf{W}_{\textrm{hh}} \in \mathbb{R}^{h \times h}$, we apply the chain rule

giving

$$\begin{aligned}
\frac{\partial L}{\partial \mathbf{W}_{\textrm{hx}}} &= \sum_{t=1}^T \textrm{prod}\left(\frac{\partial L}{\partial \mathbf{h}_t}, \frac{\partial \mathbf{h}_t}{\partial \mathbf{W}_{\textrm{hx}}}\right) = \sum_{t=1}^T \frac{\partial L}{\partial \mathbf{h}_t} \mathbf{x}_t^\top,\\
\frac{\partial L}{\partial \mathbf{W}_{\textrm{hh}}} &= \sum_{t=1}^T \textrm{prod}\left(\frac{\partial L}{\partial \mathbf{h}_t}, \frac{\partial \mathbf{h}_t}{\partial \mathbf{W}_{\textrm{hh}}}\right) = \sum_{t=1}^T \frac{\partial L}{\partial \mathbf{h}_t} \mathbf{h}_{t-1}^\top,
\end{aligned} \tag{9.7.16}$$

where $\partial L/\partial \mathbf{h}_t$ which is recurrently computed by (9.7.13) and (9.7.14) is the key quantity that affects the numerical stability.

Since backpropagation through time is the application of backpropagation in RNNs, as we have explained in Section 5.3, training RNNs alternates forward propagation with backpropagation through time. Moreover, backpropagation through time computes and stores the above gradients in turn. Specifically, stored intermediate values are reused to avoid duplicate calculations, such as storing $\partial L/\partial \mathbf{h}_t$ to be used in computation of both $\partial L / \partial \mathbf{W}_{\textrm{hx}}$ and $\partial L / \partial \mathbf{W}_{\textrm{hh}}$.

### 9.7.3 Summary

Backpropagation through time is merely an application of backpropagation to sequence models with a hidden state. Truncation, such as regular or randomized, is needed for computational convenience and numerical stability. High powers of matrices can lead to divergent or vanishing eigenvalues. This manifests itself in the form of exploding or vanishing gradients. For efficient computation, intermediate values are cached during backpropagation through time.

### 9.7.4 Exercises

1. Assume that we have a symmetric matrix $\mathbf{M} \in \mathbb{R}^{n \times n}$ with eigenvalues $\lambda_i$ whose corresponding eigenvectors are $\mathbf{v}_i$ ($i = 1, \ldots, n$). Without loss of generality, assume that they are ordered in the order $|\lambda_i| \geq |\lambda_{i+1}|$.
2. Show that $\mathbf{M}^k$ has eigenvalues $\lambda_i^k$.
3. Prove that for a random vector $\mathbf{x} \in \mathbb{R}^n$, with high probability $\mathbf{M}^k \mathbf{x}$ will be very much aligned with the eigenvector $\mathbf{v}_1$ of $\mathbf{M}$. Formalize this statement.
4. What does the above result mean for gradients in RNNs?
5. Besides gradient clipping, can you think of any other methods to cope with gradient explosion in recurrent neural networks?

Discussions[144].

# 10 Modern Recurrent Neural Networks

The previous chapter introduced the key ideas behind recurrent neural networks (RNNs). However, just as with convolutional neural networks, there has been a tremendous amount of innovation in RNN architectures, culminating in several complex designs that have proven successful in practice. In particular, the most popular designs feature mechanisms for mitigating the notorious numerical instability faced by RNNs, as typified by vanishing and exploding gradients. Recall that in Chapter 9 we dealt with exploding gradients by applying a blunt gradient clipping heuristic. Despite the efficacy of this hack, it leaves open the problem of vanishing gradients.

In this chapter, we introduce the key ideas behind the most successful RNN architectures for sequences, which stem from two papers. The first, *Long Short-Term Memory* (Hochreiter and Schmidhuber, 1997), introduces the *memory cell*, a unit of computation that replaces traditional nodes in the hidden layer of a network. With these memory cells, networks are able to overcome difficulties with training encountered by earlier recurrent networks. Intuitively, the memory cell avoids the vanishing gradient problem by keeping values in each memory cell's internal state cascading along a recurrent edge with weight 1 across many successive time steps. A set of multiplicative gates help the network to determine not only the inputs to allow into the memory state, but when the content of the memory state should influence the model's output.

The second paper, *Bidirectional Recurrent Neural Networks* (Schuster and Paliwal, 1997), introduces an architecture in which information from both the future (subsequent time steps) and the past (preceding time steps) are used to determine the output at any point in the sequence. This is in contrast to previous networks, in which only past input can affect the output. Bidirectional RNNs have become a mainstay for sequence labeling tasks in natural language processing, among a myriad of other tasks. Fortunately, the two innovations are not mutually exclusive, and have been successfully combined for phoneme classification (Graves and Schmidhuber, 2005) and handwriting recognition (Graves *et al.*, 2008).

The first sections in this chapter will explain the LSTM architecture, a lighter-weight version called the gated recurrent unit (GRU), the key ideas behind bidirectional RNNs and a brief explanation of how RNN layers are stacked together to form deep RNNs. Subsequently, we will explore the application of RNNs in sequence-to-sequence tasks, introducing machine translation along with key ideas such as *encoder–decoder* architectures and *beam search*.

# 10.1 Long Short-Term Memory (LSTM)

Shortly after the first Elman-style RNNs were trained using backpropagation (Elman, 1990), the problems of learning long-term dependencies (owing to vanishing and exploding gradients) became salient, with Bengio and Hochreiter discussing the problem (Bengio *et al.*, 1994, Hochreiter *et al.*, 2001). Hochreiter had articulated this problem as early as 1991 in his Master's thesis, although the results were not widely known because the thesis was written in German. While gradient clipping helps with exploding gradients, handling vanishing gradients appears to require a more elaborate solution. One of the first and most successful techniques for addressing vanishing gradients came in the form of the long short-term memory (LSTM) model due to Hochreiter and Schmidhuber (1997). LSTMs resemble standard recurrent neural networks but here each ordinary recurrent node is replaced by a *memory cell*. Each memory cell contains an *internal state*, i.e., a node with a self-connected recurrent edge of fixed weight 1, ensuring that the gradient can pass across many time steps without vanishing or exploding.

The term "long short-term memory" comes from the following intuition. Simple recurrent neural networks have *long-term memory* in the form of weights. The weights change slowly during training, encoding general knowledge about the data. They also have *short-term memory* in the form of ephemeral activations, which pass from each node to successive nodes. The LSTM model introduces an intermediate type of storage via the memory cell. A memory cell is a composite unit, built from simpler nodes in a specific connectivity pattern, with the novel inclusion of multiplicative nodes.

```
import torch
from torch import nn
from d2l import torch as d2l
```

## 10.1.1 Gated Memory Cell

Each memory cell is equipped with an *internal state* and a number of multiplicative gates that determine whether (i) a given input should impact the internal state (the *input gate*), (ii) the internal state should be flushed to 0 (the *forget gate*), and (iii) the internal state of a given neuron should be allowed to impact the cell's output (the *output* gate).

### Gated Hidden State

The key distinction between vanilla RNNs and LSTMs is that the latter support gating of the hidden state. This means that we have dedicated mechanisms for when a hidden state should be *updated* and also for when it should be *reset*. These mechanisms are learned and they address the concerns listed above. For instance, if the first token is of great importance we will learn not to update the hidden state after the first observation. Likewise, we will learn to skip irrelevant temporary observations. Last, we will learn to reset the latent state whenever needed. We discuss this in detail below.

### Input Gate, Forget Gate, and Output Gate

The data feeding into the LSTM gates are the input at the current time step and the hidden state of the previous time step, as illustrated in Fig. 10.1.1. Three fully connected layers with sigmoid activation functions compute the values of the input, forget, and output gates. As a result of the sigmoid activation, all values of the three gates are in the range of $(0, 1)$. Additionally, we require an *input node*, typically computed with a *tanh* activation function. Intuitively, the *input gate* determines how much of the input node's value should be added to the current memory cell internal state. The *forget gate* determines whether to keep the current value of the memory or flush it. And the *output gate* determines whether the memory cell should influence the output at the current time step.

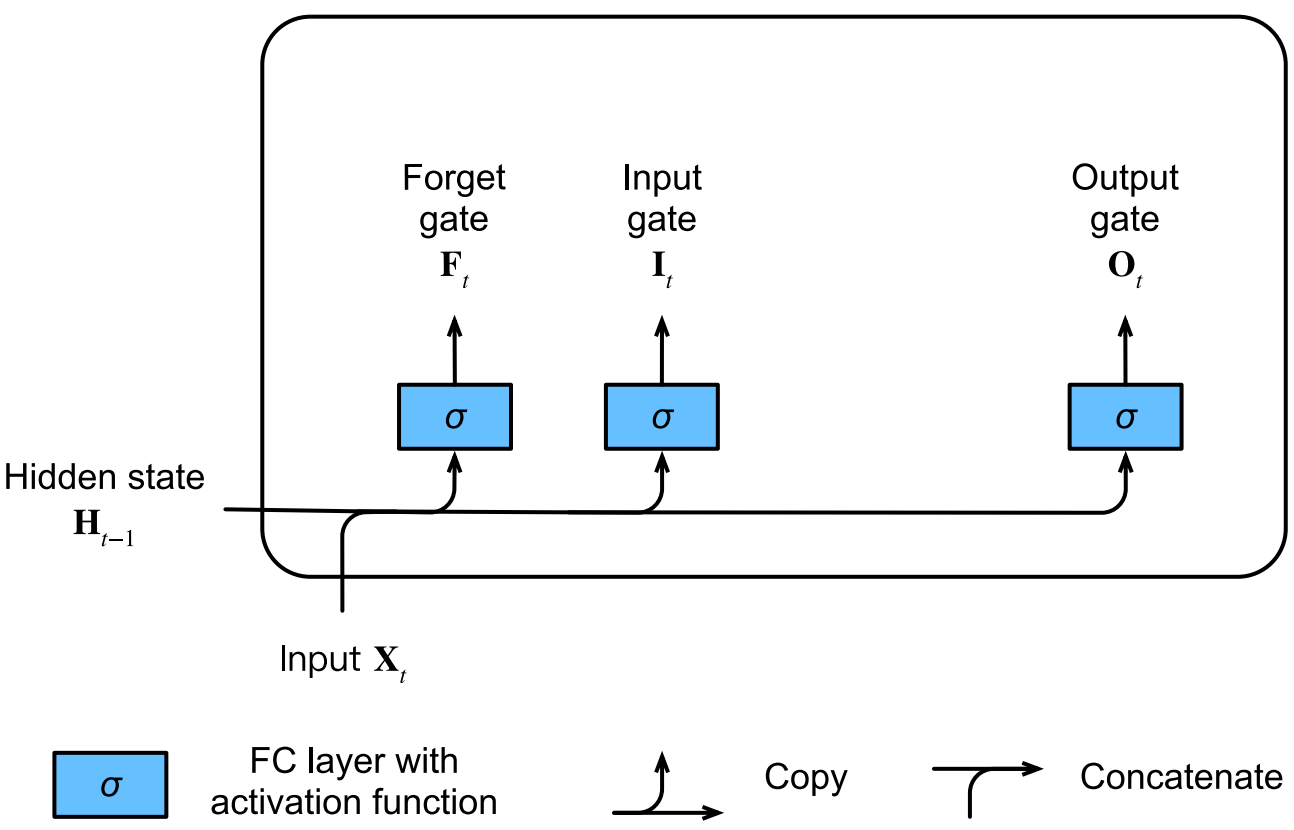

Fig. 10.1.1 Computing the input gate, the forget gate, and the output gate in an LSTM model.

Mathematically, suppose that there are $h$ hidden units, the batch size is $n$, and the number of inputs is $d$. Thus, the input is $\mathbf{X}_t \in \mathbb{R}^{n\times d}$ and the hidden state of the previous time step is $\mathbf{H}_{t-1} \in \mathbb{R}^{n\times h}$. Correspondingly, the gates at time step $t$ are defined as follows: the input gate is $\mathbf{I}_t \in \mathbb{R}^{n\times h}$, the forget gate is $\mathbf{F}_t \in \mathbb{R}^{n\times h}$, and the output gate is $\mathbf{O}_t \in \mathbb{R}^{n\times h}$. They are calculated as follows:

$$
\begin{aligned}
\mathbf{I}_t &= \sigma(\mathbf{X}_t \mathbf{W}_{\textrm{xi}} + \mathbf{H}_{t-1} \mathbf{W}_{\textrm{hi}} + \mathbf{b}_\textrm{i}),\\
\mathbf{F}_t &= \sigma(\mathbf{X}_t \mathbf{W}_{\textrm{xf}} + \mathbf{H}_{t-1} \mathbf{W}_{\textrm{hf}} + \mathbf{b}_\textrm{f}),\\
\mathbf{O}_t &= \sigma(\mathbf{X}_t \mathbf{W}_{\textrm{xo}} + \mathbf{H}_{t-1} \mathbf{W}_{\textrm{ho}} + \mathbf{b}_\textrm{o}),
\end{aligned}
\tag{10.1.1}
$$

where $\mathbf{W}_{\textrm{xi}}, \mathbf{W}_{\textrm{xf}}, \mathbf{W}_{\textrm{xo}} \in \mathbb{R}^{d\times h}$ and $\mathbf{W}_{\textrm{hi}}, \mathbf{W}_{\textrm{hf}}, \mathbf{W}_{\textrm{ho}} \in \mathbb{R}^{h\times h}$ are weight parameters and $\mathbf{b}_\textrm{i}, \mathbf{b}_\textrm{f}, \mathbf{b}_\textrm{o} \in \mathbb{R}^{1\times h}$ are bias parameters. Note that broadcasting (see Section 2.1.4) is triggered during the summation. We use sigmoid functions (as introduced in Section 5.1) to map the input values to the interval $(0, 1)$.

### Input Node

Next we design the memory cell. Since we have not specified the action of the various gates yet, we first introduce the *input node* $\tilde{\mathbf{C}}_t \in \mathbb{R}^{n\times h}$. Its computation is similar to that of the three gates described above, but uses a $\tanh$ function with a value range for $(-1, 1)$ as the activation function. This leads to the following equation at time step $t$:

$$\tilde{\mathbf{C}}_t = \tanh(\mathbf{X}_t\mathbf{W}_{\textrm{xc}} + \mathbf{H}_{t-1}\mathbf{W}_{\textrm{hc}} + \mathbf{b}_\textrm{c}), \tag{10.1.2}$$

where $\mathbf{W}_{\textrm{xc}} \in \mathbb{R}^{d\times h}$ and $\mathbf{W}_{\textrm{hc}} \in \mathbb{R}^{h\times h}$ are weight parameters and $\mathbf{b}_\textrm{c} \in \mathbb{R}^{1\times h}$ is a bias parameter.

A quick illustration of the input node is shown in Fig. 10.1.2.

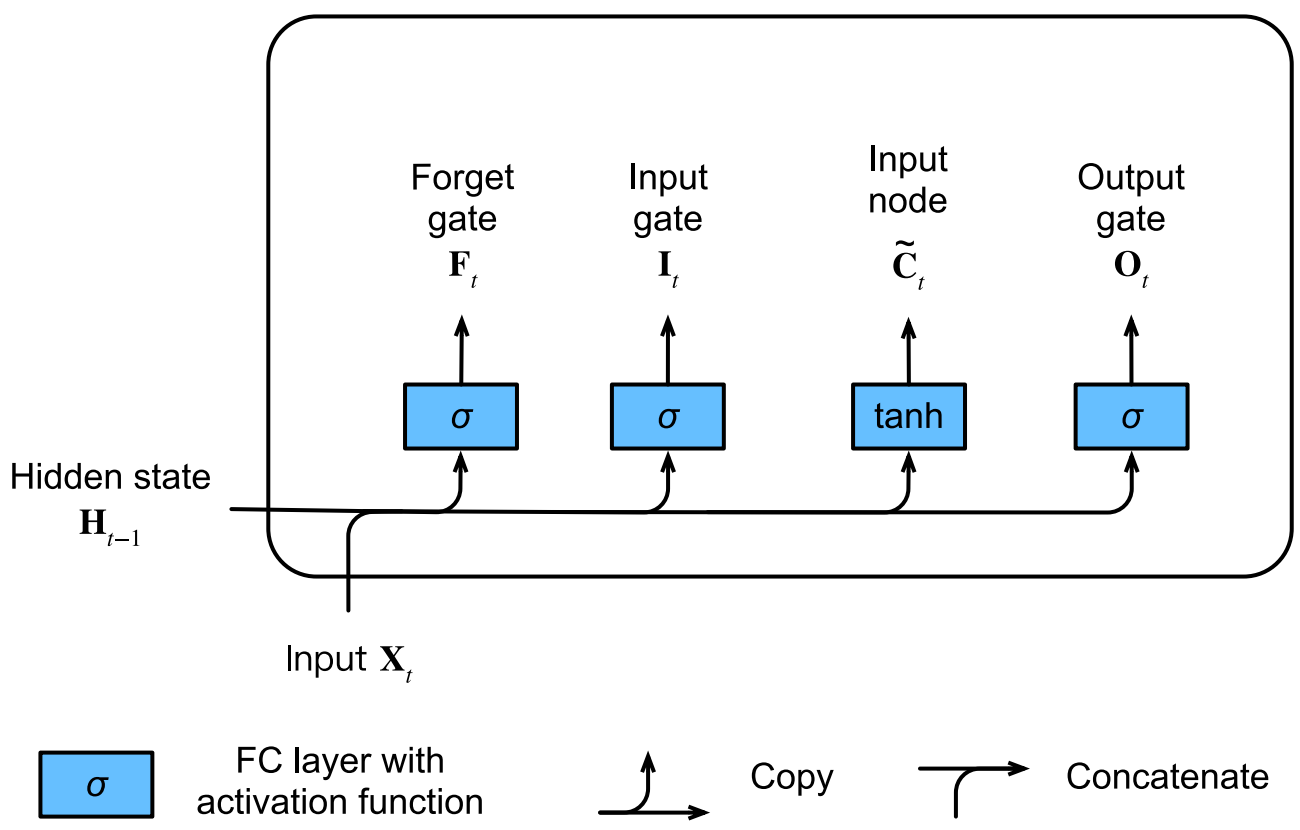

Fig. 10.1.2 Computing the input node in an LSTM model.

### Memory Cell Internal State

In LSTMs, the input gate $\mathbf{I}_t$ governs how much we take new data into account via $\tilde{\mathbf{C}}_t$ and the forget gate $\mathbf{F}_t$ addresses how much of the old cell internal state $\mathbf{C}_{t-1} \in \mathbb{R}^{n\times h}$ we retain. Using the Hadamard (elementwise) product operator $\odot$ we arrive at the following update equation:

$$\mathbf{C}_t = \mathbf{F}_t \odot \mathbf{C}_{t-1} + \mathbf{I}_t \odot \tilde{\mathbf{C}}_t. \tag{10.1.3}$$

If the forget gate is always 1 and the input gate is always 0, the memory cell internal state $\mathbf{C}_{t-1}$ will remain constant forever, passing unchanged to each subsequent time step. However, input gates and forget gates give the model the flexibility of being able to learn when to keep this value unchanged and when to perturb it in response to subsequent inputs. In practice, this design alleviates the vanishing gradient problem, resulting in models that are much easier to train, especially when facing datasets with long sequence lengths.

We thus arrive at the flow diagram in Fig. 10.1.3.

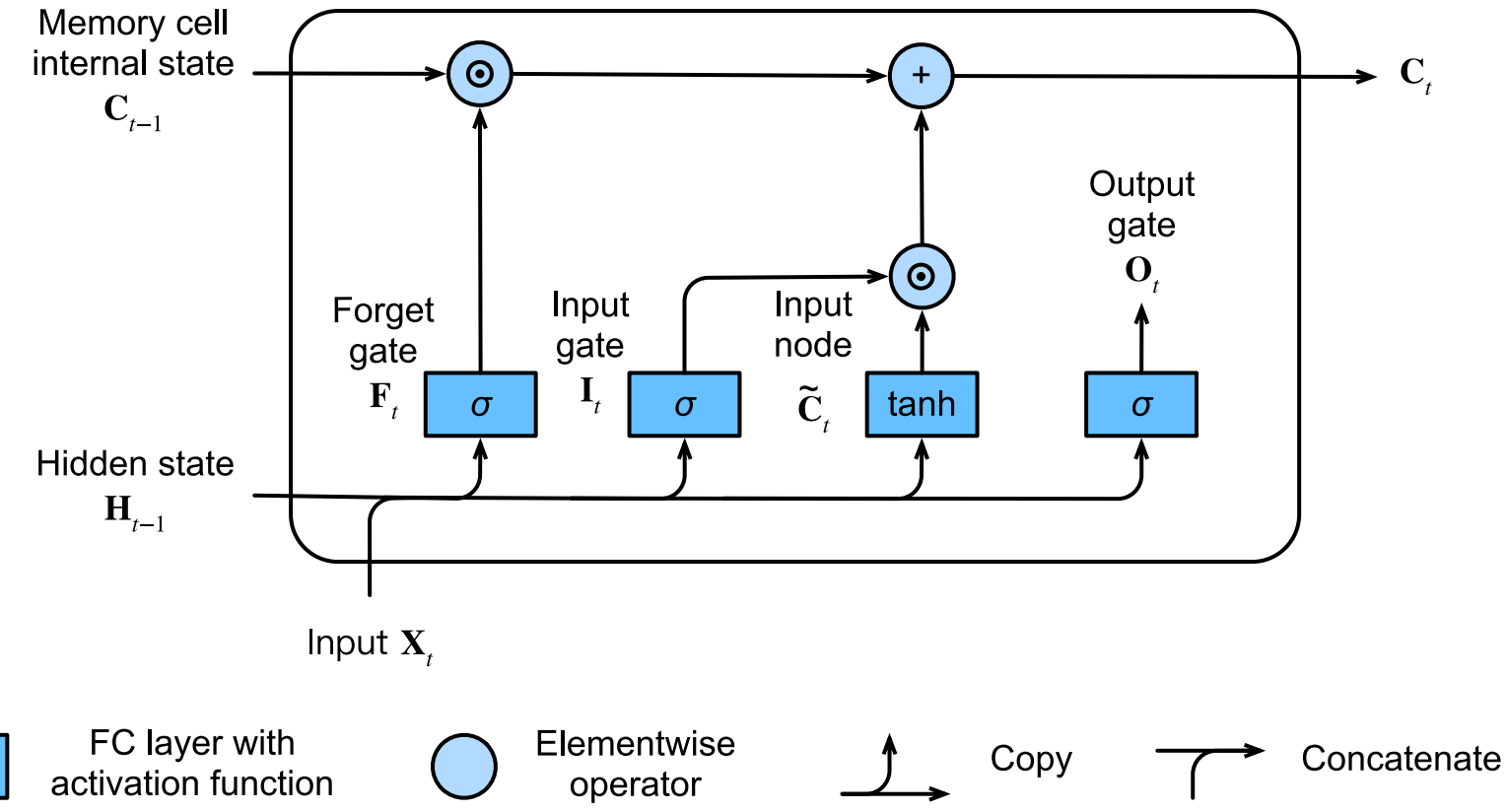

Fig. 10.1.3 Computing the memory cell internal state in an LSTM model.

### Hidden State

Last, we need to define how to compute the output of the memory cell, i.e., the hidden state $\mathbf{H}_t \in \mathbb{R}^{n \times h}$, as seen by other layers. This is where the output gate comes into play. In LSTMs, we first apply $\tanh$ to the memory cell internal state and then apply another point-wise multiplication, this time with the output gate. This ensures that the values of $\mathbf{H}_t$ are always in the interval $(-1, 1)$:

$$\mathbf{H}_t = \mathbf{O}_t \odot \tanh(\mathbf{C}_t). \tag{10.1.4}$$

Whenever the output gate is close to 1, we allow the memory cell internal state to impact the subsequent layers uninhibited, whereas for output gate values close to 0, we prevent the current memory from impacting other layers of the network at the current time step. Note that a memory cell can accrue information across many time steps without impacting the rest of the network (as long as the output gate takes values close to 0), and then suddenly impact the network at a subsequent time step as soon as the output gate flips from values close to 0 to values close to 1. Fig. 10.1.4 has a graphical illustration of the data flow.

## 10.1.2 Implementation from Scratch

Now let's implement an LSTM from scratch. As same as the experiments in Section 9.5, we first load *The Time Machine* dataset.

### Initializing Model Parameters

Next, we need to define and initialize the model parameters. As previously, the hyperparameter `num_hiddens` dictates the number of hidden units. We initialize weights following a Gaussian distribution with 0.01 standard deviation, and we set the biases to 0.

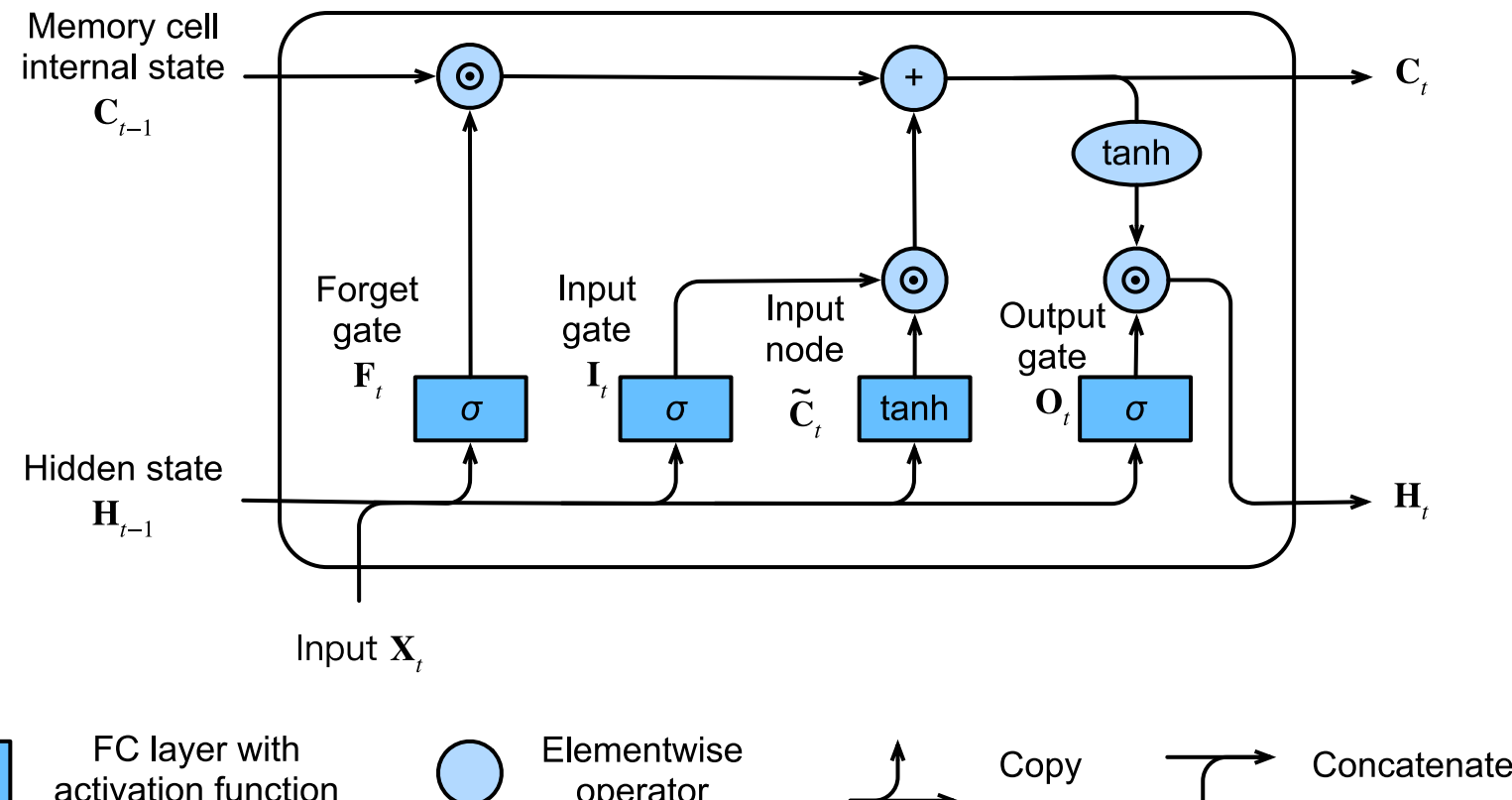

Fig. 10.1.4 Computing the hidden state in an LSTM model.

```
class LSTMScratch(d2l.Module):
    def __init__(self, num_inputs, num_hiddens, sigma=0.01):
        super().__init__()
        self.save_hyperparameters()

        init_weight = lambda *shape: nn.Parameter(torch.randn(*shape) * sigma)
        triple = lambda: (init_weight(num_inputs, num_hiddens),
                          init_weight(num_hiddens, num_hiddens),
                          nn.Parameter(torch.zeros(num_hiddens)))
        self.W_xi, self.W_hi, self.b_i = triple()  # Input gate
        self.W_xf, self.W_hf, self.b_f = triple()  # Forget gate
        self.W_xo, self.W_ho, self.b_o = triple()  # Output gate
        self.W_xc, self.W_hc, self.b_c = triple()  # Input node
```

The actual model is defined as described above, consisting of three gates and an input node. Note that only the hidden state is passed to the output layer.

```
@d2l.add_to_class(LSTMScratch)
def forward(self, inputs, H_C=None):
    if H_C is None:
        # Initial state with shape: (batch_size, num_hiddens)
        H = torch.zeros((inputs.shape[1], self.num_hiddens),
                      device=inputs.device)
        C = torch.zeros((inputs.shape[1], self.num_hiddens),
                      device=inputs.device)
    else:
        H, C = H_C
    outputs = []
    for X in inputs:
        I = torch.sigmoid(torch.matmul(X, self.W_xi) +
                        torch.matmul(H, self.W_hi) + self.b_i)
        F = torch.sigmoid(torch.matmul(X, self.W_xf) +
                        torch.matmul(H, self.W_hf) + self.b_f)
        O = torch.sigmoid(torch.matmul(X, self.W_xo) +
                        torch.matmul(H, self.W_ho) + self.b_o)
```

```
        C_tilde = torch.tanh(torch.matmul(X, self.W_xc) +
                           torch.matmul(H, self.W_hc) + self.b_c)
        C = F * C + I * C_tilde
        H = O * torch.tanh(C)
        outputs.append(H)
    return outputs, (H, C)
```

### Training and Prediction

Let's train an LSTM model by instantiating the `RNNLMScratch` class from Section 9.5.

```
data = d2l.TimeMachine(batch_size=1024, num_steps=32)
lstm = LSTMScratch(num_inputs=len(data.vocab), num_hiddens=32)
model = d2l.RNNLMScratch(lstm, vocab_size=len(data.vocab), lr=4)
trainer = d2l.Trainer(max_epochs=50, gradient_clip_val=1, num_gpus=1)
trainer.fit(model, data)
```

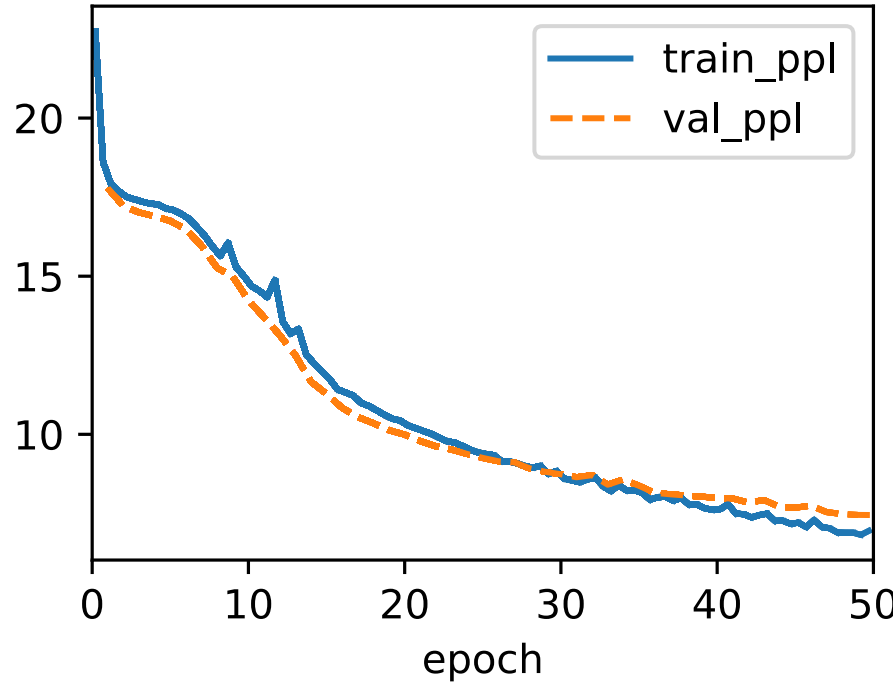

## 10.1.3 Concise Implementation

Using high-level APIs, we can directly instantiate an LSTM model. This encapsulates all the configuration details that we made explicit above. The code is significantly faster as it uses compiled operators rather than Python for many details that we spelled out before.

```
class LSTM(d2l.RNN):
    def __init__(self, num_inputs, num_hiddens):
        d2l.Module.__init__(self)
        self.save_hyperparameters()
        self.rnn = nn.LSTM(num_inputs, num_hiddens)

    def forward(self, inputs, H_C=None):
        return self.rnn(inputs, H_C)
```

```
lstm = LSTM(num_inputs=len(data.vocab), num_hiddens=32)
model = d2l.RNNLM(lstm, vocab_size=len(data.vocab), lr=4)
trainer.fit(model, data)
```

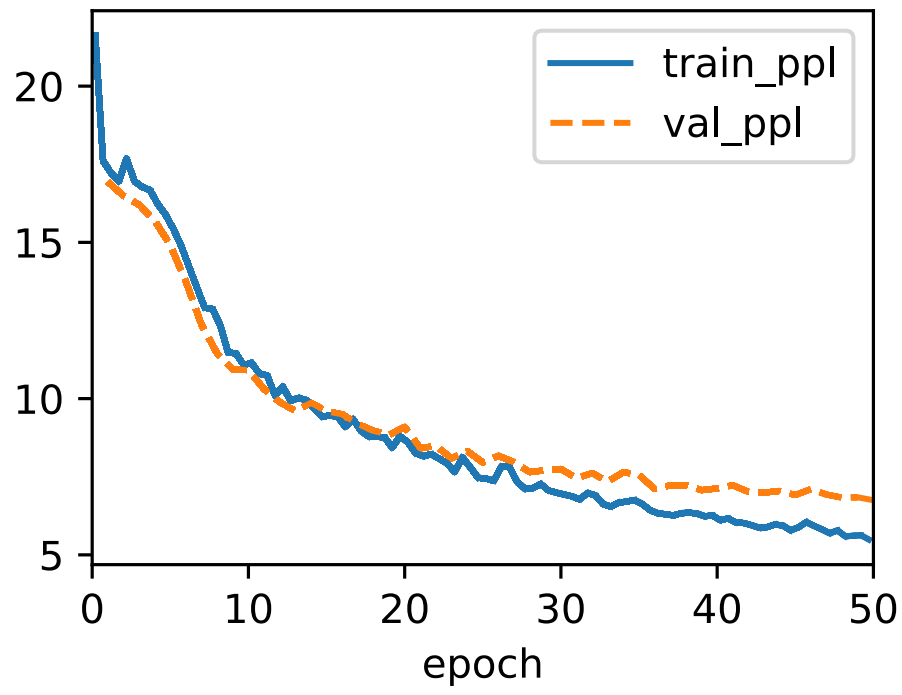

```
model.predict('it has', 20, data.vocab, d2l.try_gpu())
```

```
'it has the the the the the'
```

LSTMs are the prototypical latent variable autoregressive model with nontrivial state control. Many variants thereof have been proposed over the years, e.g., multiple layers, residual connections, different types of regularization. However, training LSTMs and other sequence models (such as GRUs) is quite costly because of the long range dependency of the sequence. Later we will encounter alternative models such as Transformers that can be used in some cases.

### 10.1.4 Summary

While LSTMs were published in 1997, they rose to great prominence with some victories in prediction competitions in the mid-2000s, and became the dominant models for sequence learning from 2011 until the rise of Transformer models, starting in 2017. Even Transformers owe some of their key ideas to architecture design innovations introduced by the LSTM.

LSTMs have three types of gates: input gates, forget gates, and output gates that control the flow of information. The hidden layer output of LSTM includes the hidden state and the memory cell internal state. Only the hidden state is passed into the output layer while the memory cell internal state remains entirely internal. LSTMs can alleviate vanishing and exploding gradients.

### 10.1.5 Exercises

1. Adjust the hyperparameters and analyze their influence on running time, perplexity, and the output sequence.
2. How would you need to change the model to generate proper words rather than just sequences of characters?
3. Compare the computational cost for GRUs, LSTMs, and regular RNNs for a given hidden dimension. Pay special attention to the training and inference cost.
4. Since the candidate memory cell ensures that the value range is between $-1$ and $1$ by using the $\tanh$ function, why does the hidden state need to use the $\tanh$ function again to ensure that the output value range is between $-1$ and $1$?
5. Implement an LSTM model for time series prediction rather than character sequence prediction.

Discussions[145].

# 10.2 Gated Recurrent Units (GRU)

As RNNs and particularly the LSTM architecture (Section 10.1) rapidly gained popularity during the 2010s, a number of researchers began to experiment with simplified architectures in hopes of retaining the key idea of incorporating an internal state and multiplicative gating mechanisms but with the aim of speeding up computation. The gated recurrent unit (GRU) (Cho *et al.*, 2014) offered a streamlined version of the LSTM memory cell that often achieves comparable performance but with the advantage of being faster to compute (Chung *et al.*, 2014).

```
import torch
from torch import nn
from d2l import torch as d2l
```

## 10.2.1 Reset Gate and Update Gate

Here, the LSTM's three gates are replaced by two: the *reset gate* and the *update gate*. As with LSTMs, these gates are given sigmoid activations, forcing their values to lie in the interval $(0, 1)$. Intuitively, the reset gate controls how much of the previous state we might still want to remember. Likewise, an update gate would allow us to control how much of the new state is just a copy of the old one. Fig. 10.2.1 illustrates the inputs for both the reset and update gates in a GRU, given the input of the current time step and the hidden state of the previous time step. The outputs of the gates are given by two fully connected layers with a sigmoid activation function.

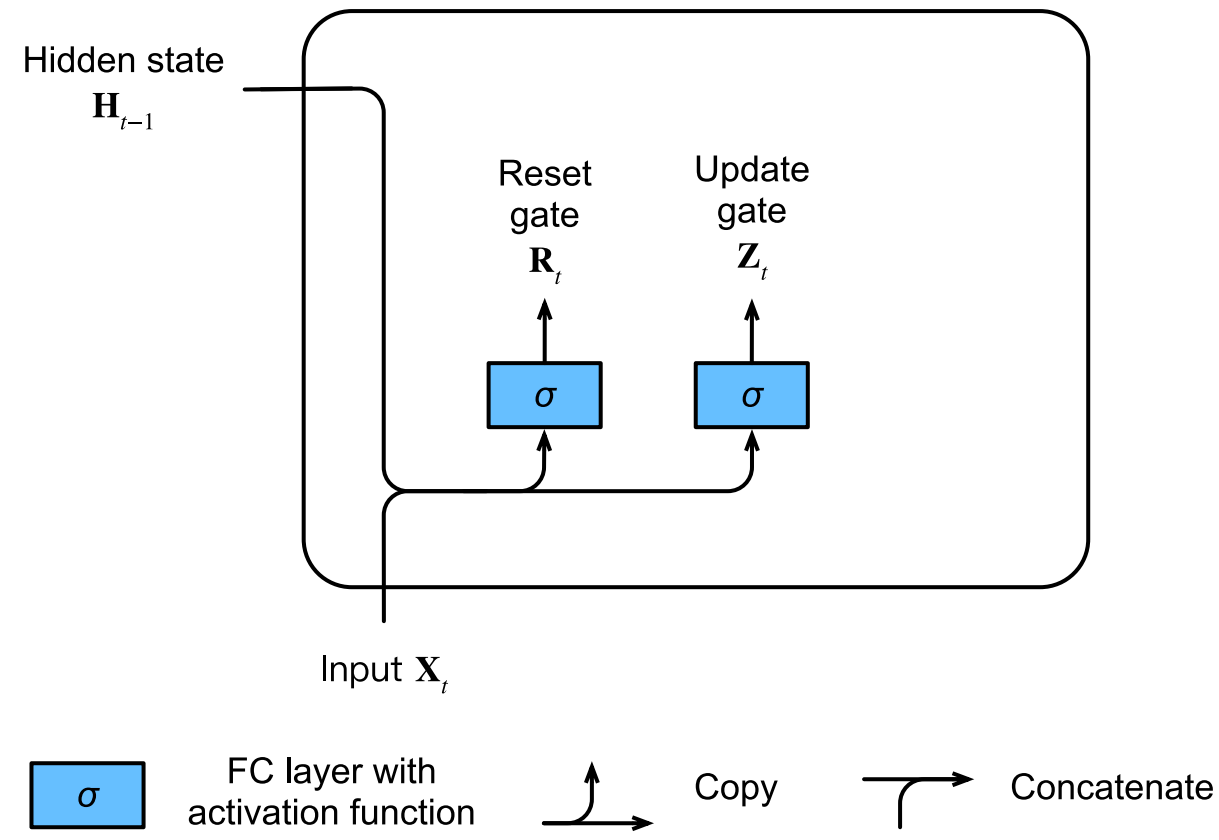

Fig. 10.2.1 Computing the reset gate and the update gate in a GRU model.

Mathematically, for a given time step $t$, suppose that the input is a minibatch $\mathbf{X}_t \in \mathbb{R}^{n\times d}$ (number of examples $= n$; number of inputs $= d$) and the hidden state of the previous time step is $\mathbf{H}_{t-1} \in \mathbb{R}^{n\times h}$ (number of hidden units $= h$). Then the reset gate $\mathbf{R}_t \in \mathbb{R}^{n\times h}$ and update gate $\mathbf{Z}_t \in \mathbb{R}^{n\times h}$ are computed as follows:

$$\begin{aligned} \mathbf{R}_t &= \sigma(\mathbf{X}_t \mathbf{W}_{\textrm{xr}} + \mathbf{H}_{t-1} \mathbf{W}_{\textrm{hr}} + \mathbf{b}_\textrm{r}),\\ \mathbf{Z}_t &= \sigma(\mathbf{X}_t \mathbf{W}_{\textrm{xz}} + \mathbf{H}_{t-1} \mathbf{W}_{\textrm{hz}} + \mathbf{b}_\textrm{z}), \end{aligned} \tag{10.2.1}$$

where $\mathbf{W}_{\textrm{xr}}, \mathbf{W}_{\textrm{xz}} \in \mathbb{R}^{d\times h}$ and $\mathbf{W}_{\textrm{hr}}, \mathbf{W}_{\textrm{hz}} \in \mathbb{R}^{h\times h}$ are weight parameters and $\mathbf{b}_\textrm{r}, \mathbf{b}_\textrm{z} \in \mathbb{R}^{1\times h}$ are bias parameters.

## 10.2.2 Candidate Hidden State

Next, we integrate the reset gate $\mathbf{R}_t$ with the regular updating mechanism in (9.4.5), leading to the following *candidate hidden state* $\tilde{\mathbf{H}}_t \in \mathbb{R}^{n\times h}$ at time step $t$:

$$\tilde{\mathbf{H}}_t = \tanh(\mathbf{X}_t \mathbf{W}_{\textrm{xh}} + \left(\mathbf{R}_t \odot \mathbf{H}_{t-1}\right) \mathbf{W}_{\textrm{hh}} + \mathbf{b}_\textrm{h}), \tag{10.2.2}$$

where $\mathbf{W}_{\textrm{xh}} \in \mathbb{R}^{d\times h}$ and $\mathbf{W}_{\textrm{hh}} \in \mathbb{R}^{h\times h}$ are weight parameters, $\mathbf{b}_\textrm{h} \in \mathbb{R}^{1\times h}$ is the bias, and the symbol $\odot$ is the Hadamard (elementwise) product operator. Here we use a tanh activation function.

The result is a *candidate*, since we still need to incorporate the action of the update gate. Comparing with (9.4.5), the influence of the previous states can now be reduced with the elementwise multiplication of $\mathbf{R}_t$ and $\mathbf{H}_{t-1}$ in (10.2.2). Whenever the entries in the reset gate $\mathbf{R}_t$ are close to 1, we recover a vanilla RNN such as that in (9.4.5). For all entries of the reset gate $\mathbf{R}_t$ that are close to 0, the candidate hidden state is the result of an MLP with $\mathbf{X}_t$ as input. Any pre-existing hidden state is thus *reset* to defaults.

Fig. 10.2.2 illustrates the computational flow after applying the reset gate.

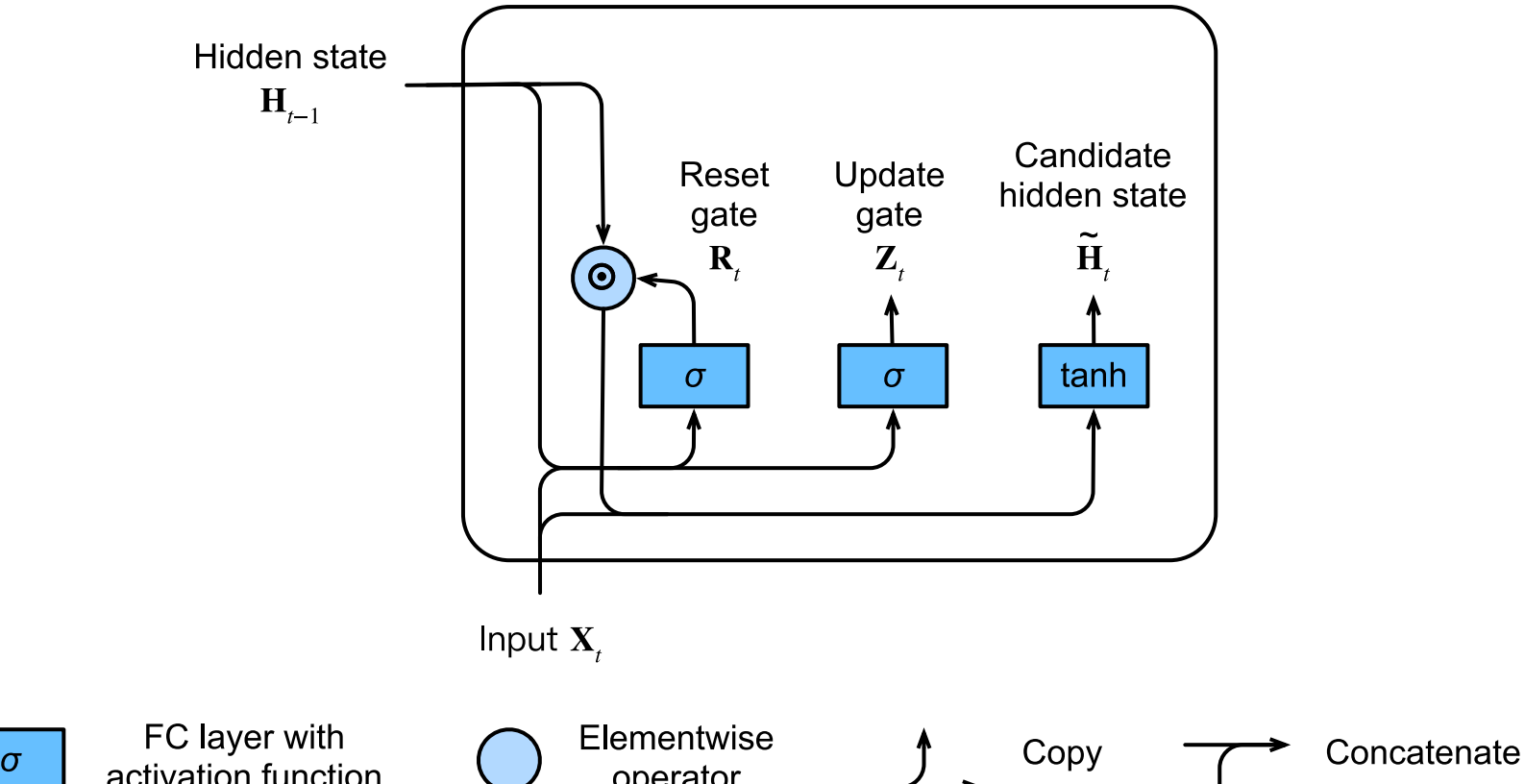

Fig. 10.2.2 Computing the candidate hidden state in a GRU model.

### 10.2.3 Hidden State

Finally, we need to incorporate the effect of the update gate $\mathbf{Z}_t$. This determines the extent to which the new hidden state $\mathbf{H}_t \in \mathbb{R}^{n \times h}$ matches the old state $\mathbf{H}_{t-1}$ compared with how much it resembles the new candidate state $\tilde{\mathbf{H}}_t$. The update gate $\mathbf{Z}_t$ can be used for this purpose, simply by taking elementwise convex combinations of $\mathbf{H}_{t-1}$ and $\tilde{\mathbf{H}}_t$. This leads to the final update equation for the GRU:

$$\mathbf{H}_t = \mathbf{Z}_t \odot \mathbf{H}_{t-1} + (1 - \mathbf{Z}_t) \odot \tilde{\mathbf{H}}_t. \tag{10.2.3}$$

Whenever the update gate $\mathbf{Z}_t$ is close to 1, we simply retain the old state. In this case the information from $\mathbf{X}_t$ is ignored, effectively skipping time step $t$ in the dependency chain. By contrast, whenever $\mathbf{Z}_t$ is close to 0, the new latent state $\mathbf{H}_t$ approaches the candidate latent state $\tilde{\mathbf{H}}_t$. Fig. 10.2.3 shows the computational flow after the update gate is in action.

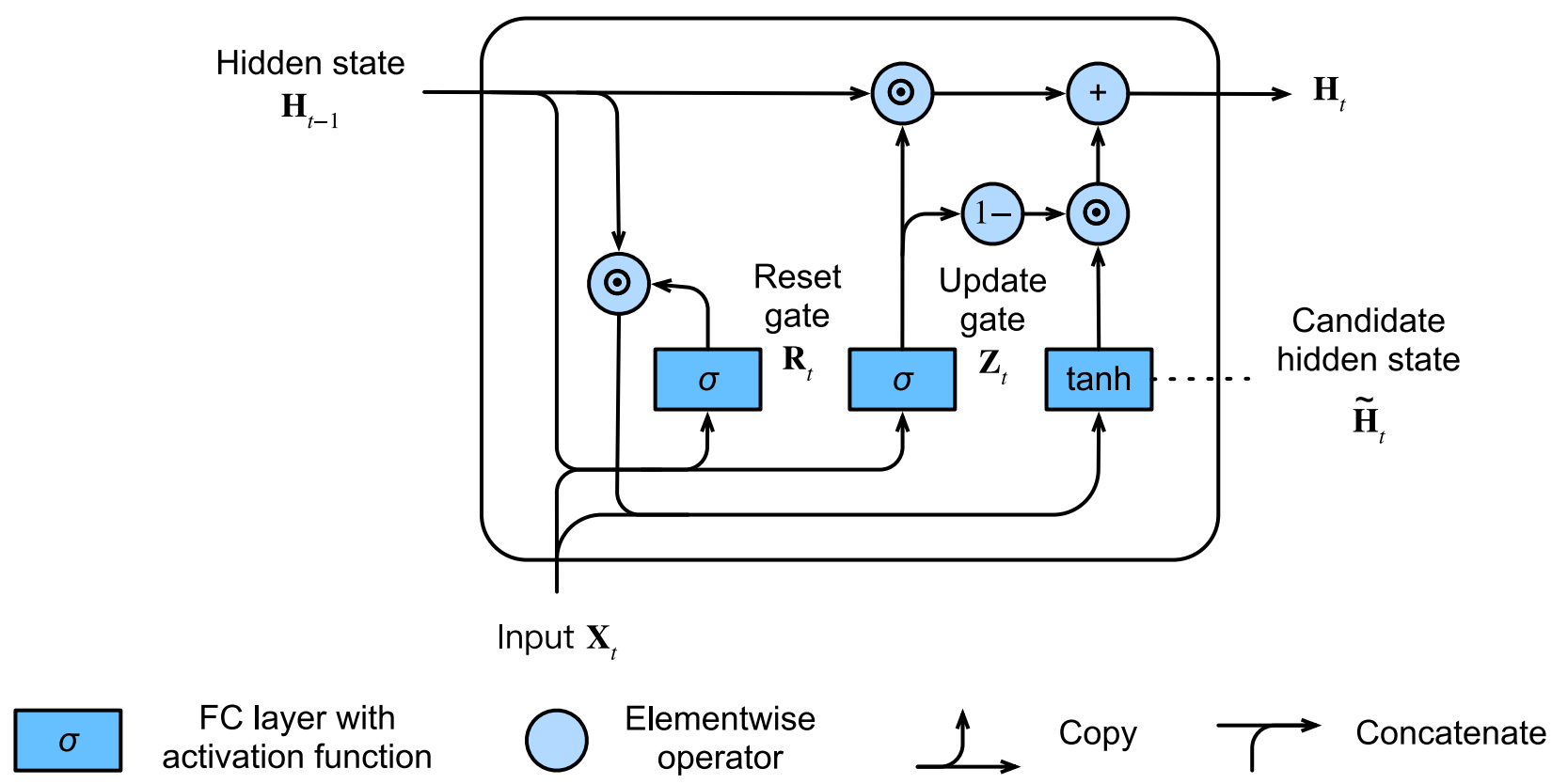

Fig. 10.2.3 Computing the hidden state in a GRU model.

In summary, GRUs have the following two distinguishing features:

- Reset gates help capture short-term dependencies in sequences.
- Update gates help capture long-term dependencies in sequences.

## 10.2.4 Implementation from Scratch

To gain a better understanding of the GRU model, let's implement it from scratch.

### Initializing Model Parameters

The first step is to initialize the model parameters. We draw the weights from a Gaussian distribution with standard deviation to be `sigma` and set the bias to 0. The hyperparameter `num_hiddens` defines the number of hidden units. We instantiate all weights and biases relating to the update gate, the reset gate, and the candidate hidden state.

```
class GRUScratch(d2l.Module):
    def __init__(self, num_inputs, num_hiddens, sigma=0.01):
        super().__init__()
        self.save_hyperparameters()

        init_weight = lambda *shape: nn.Parameter(torch.randn(*shape) * sigma)
        triple = lambda: (init_weight(num_inputs, num_hiddens),
                          init_weight(num_hiddens, num_hiddens),
                          nn.Parameter(torch.zeros(num_hiddens)))
        self.W_xz, self.W_hz, self.b_z = triple()  # Update gate
        self.W_xr, self.W_hr, self.b_r = triple()  # Reset gate
        self.W_xh, self.W_hh, self.b_h = triple()  # Candidate hidden state
```

### Defining the Model

Now we are ready to define the GRU forward computation. Its structure is the same as that of the basic RNN cell, except that the update equations are more complex.

```
@d2l.add_to_class(GRUScratch)
def forward(self, inputs, H=None):
    if H is None:
        # Initial state with shape: (batch_size, num_hiddens)
        H = torch.zeros((inputs.shape[1], self.num_hiddens),
                      device=inputs.device)
    outputs = []
    for X in inputs:
        Z = torch.sigmoid(torch.matmul(X, self.W_xz) +
                        torch.matmul(H, self.W_hz) + self.b_z)
        R = torch.sigmoid(torch.matmul(X, self.W_xr) +
                        torch.matmul(H, self.W_hr) + self.b_r)
        H_tilde = torch.tanh(torch.matmul(X, self.W_xh) +
                           torch.matmul(R * H, self.W_hh) + self.b_h)
```

```
        H = Z * H + (1 - Z) * H_tilde
        outputs.append(H)
    return outputs, H
```

### Training

Training a language model on *The Time Machine* dataset works in exactly the same manner as in Section 9.5.

```
data = d2l.TimeMachine(batch_size=1024, num_steps=32)
gru = GRUScratch(num_inputs=len(data.vocab), num_hiddens=32)
model = d2l.RNNLMScratch(gru, vocab_size=len(data.vocab), lr=4)
trainer = d2l.Trainer(max_epochs=50, gradient_clip_val=1, num_gpus=1)
trainer.fit(model, data)
```

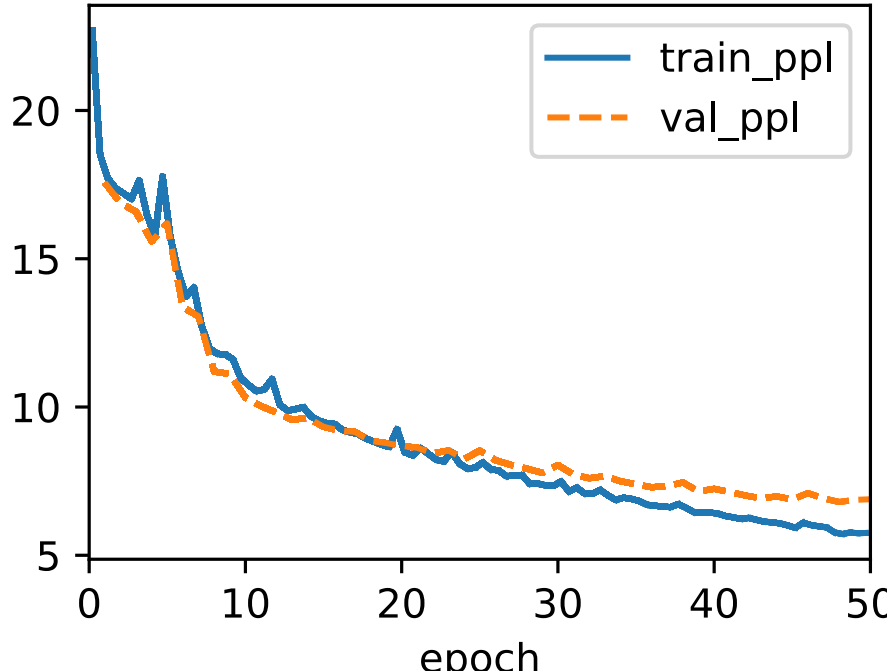

## 10.2.5 Concise Implementation

In high-level APIs, we can directly instantiate a GRU model. This encapsulates all the configuration detail that we made explicit above.

```
class GRU(d2l.RNN):
    def __init__(self, num_inputs, num_hiddens):
        d2l.Module.__init__(self)
        self.save_hyperparameters()
        self.rnn = nn.GRU(num_inputs, num_hiddens)
```

The code is significantly faster in training as it uses compiled operators rather than Python.

```
gru = GRU(num_inputs=len(data.vocab), num_hiddens=32)
model = d2l.RNNLM(gru, vocab_size=len(data.vocab), lr=4)
trainer.fit(model, data)
```

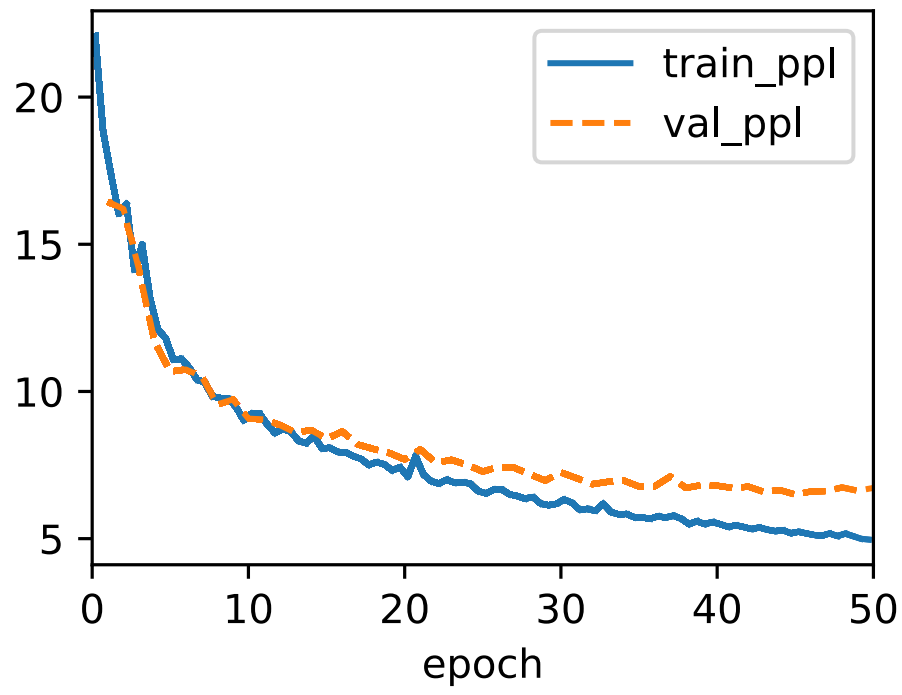

After training, we print out the perplexity on the training set and the predicted sequence following the provided prefix.

```
model.predict('it has', 20, data.vocab, d2l.try_gpu())
```

```
'it has of the fore and the'
```

## 10.2.6 Summary

Compared with LSTMs, GRUs achieve similar performance but tend to be lighter computationally. Generally, compared with simple RNNs, gated RNNS, just like LSTMs and GRUs, can better capture dependencies for sequences with large time step distances. GRUs contain basic RNNs as their extreme case whenever the reset gate is switched on. They can also skip subsequences by turning on the update gate.

## 10.2.7 Exercises

1. Assume that we only want to use the input at time step $t'$ to predict the output at time step $t > t'$. What are the best values for the reset and update gates for each time step?
2. Adjust the hyperparameters and analyze their influence on running time, perplexity, and the output sequence.
3. Compare runtime, perplexity, and the output strings for `rnn.RNN` and `rnn.GRU` implementations with each other.
4. What happens if you implement only parts of a GRU, e.g., with only a reset gate or only an update gate?

Discussions[146].

# 10.3 Deep Recurrent Neural Networks

Up until now, we have focused on defining networks consisting of a sequence input, a single hidden RNN layer, and an output layer. Despite having just one hidden layer between the input at any time step and the corresponding output, there is a sense in which these networks are deep. Inputs from the first time step can influence the outputs at the final time step $T$ (often 100s or 1000s of steps later). These inputs pass through $T$ applications of the recurrent layer before reaching the final output. However, we often also wish to retain the ability to express complex relationships between the inputs at a given time step and the outputs at that same time step. Thus we often construct RNNs that are deep not only in the time direction but also in the input-to-output direction. This is precisely the notion of depth that we have already encountered in our development of MLPs and deep CNNs.

The standard method for building this sort of deep RNN is strikingly simple: we stack the RNNs on top of each other. Given a sequence of length $T$, the first RNN produces a sequence of outputs, also of length $T$. These, in turn, constitute the inputs to the next RNN layer. In this short section, we illustrate this design pattern and present a simple example for how to code up such stacked RNNs. Below, in Fig. 10.3.1, we illustrate a deep RNN with $L$ hidden layers. Each hidden state operates on a sequential input and produces a sequential output. Moreover, any RNN cell (white box in Fig. 10.3.1) at each time step depends on both the same layer's value at the previous time step and the previous layer's value at the same time step.

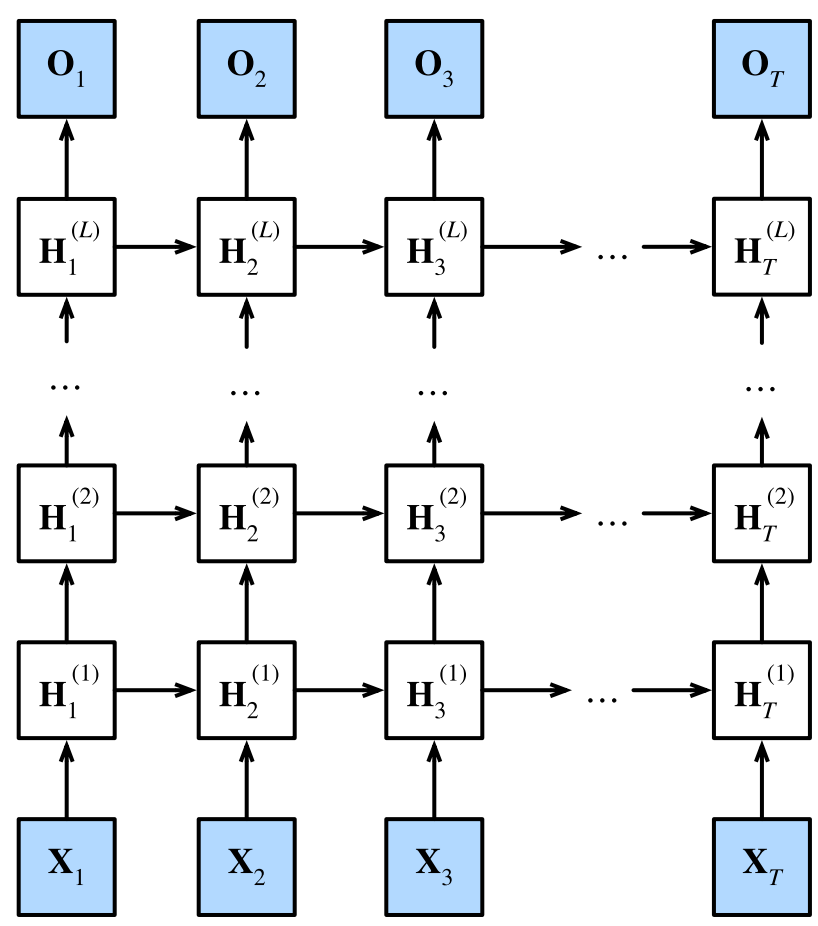

Fig. 10.3.1 Architecture of a deep RNN.

Formally, suppose that we have a minibatch input $\mathbf{X}_t \in \mathbb{R}^{n \times d}$ (number of examples $= n$; number of inputs in each example $= d$) at time step $t$. At the same time step, let the hidden state of the $l^{\text{th}}$ hidden layer ($l = 1, \ldots, L$) be $\mathbf{H}_t^{(l)} \in \mathbb{R}^{n \times h}$ (number of hidden units $= h$) and the output layer variable be $\mathbf{O}_t \in \mathbb{R}^{n \times q}$ (number of outputs: $q$). Setting $\mathbf{H}_t^{(0)} = \mathbf{X}_t$, the hidden state of the $l^{\text{th}}$ hidden layer that uses the activation function $\phi_l$ is calculated as follows:

$$\mathbf{H}_t^{(l)} = \phi_l(\mathbf{H}_t^{(l-1)} \mathbf{W}_{\text{xh}}^{(l)} + \mathbf{H}_{t-1}^{(l)} \mathbf{W}_{\text{hh}}^{(l)} + \mathbf{b}_\text{h}^{(l)}), \tag{10.3.1}$$

where the weights $\mathbf{W}_{\text{xh}}^{(l)} \in \mathbb{R}^{h \times h}$ and $\mathbf{W}_{\text{hh}}^{(l)} \in \mathbb{R}^{h \times h}$, together with the bias $\mathbf{b}_\text{h}^{(l)} \in \mathbb{R}^{1 \times h}$, are the model parameters of the $l^\text{th}$ hidden layer.

At the end, the calculation of the output layer is only based on the hidden state of the final $L^\text{th}$ hidden layer:

$$\mathbf{O}_t = \mathbf{H}_t^{(L)} \mathbf{W}_{\text{hq}} + \mathbf{b}_\text{q}, \tag{10.3.2}$$

where the weight $\mathbf{W}_{\text{hq}} \in \mathbb{R}^{h \times q}$ and the bias $\mathbf{b}_\text{q} \in \mathbb{R}^{1 \times q}$ are the model parameters of the output layer.

Just as with MLPs, the number of hidden layers $L$ and the number of hidden units $h$ are hyperparameters that we can tune. Common RNN layer widths ($h$) are in the range $(64, 2056)$, and common depths ($L$) are in the range $(1, 8)$. In addition, we can easily get a deep-gated RNN by replacing the hidden state computation in (10.3.1) with that from an LSTM or a GRU.

```
import torch
from torch import nn
from d2l import torch as d2l
```

## 10.3.1 Implementation from Scratch

To implement a multilayer RNN from scratch, we can treat each layer as an `RNNScratch` instance with its own learnable parameters.

```
class StackedRNNScratch(d2l.Module):
    def __init__(self, num_inputs, num_hiddens, num_layers, sigma=0.01):
        super().__init__()
        self.save_hyperparameters()
        self.rnns = nn.Sequential(*[d2l.RNNScratch(
            num_inputs if i==0 else num_hiddens, num_hiddens, sigma)
                                    for i in range(num_layers)])
```

The multilayer forward computation simply performs forward computation layer by layer.

```
@d2l.add_to_class(StackedRNNScratch)
def forward(self, inputs, Hs=None):
    outputs = inputs
    if Hs is None: Hs = [None] * self.num_layers
    for i in range(self.num_layers):
        outputs, Hs[i] = self.rnns[i](outputs, Hs[i])
        outputs = torch.stack(outputs, 0)
    return outputs, Hs
```

As an example, we train a deep GRU model on *The Time Machine* dataset (same as in Section 9.5). To keep things simple we set the number of layers to 2.

```
data = d2l.TimeMachine(batch_size=1024, num_steps=32)
rnn_block = StackedRNNScratch(num_inputs=len(data.vocab),
                              num_hiddens=32, num_layers=2)
model = d2l.RNNLMScratch(rnn_block, vocab_size=len(data.vocab), lr=2)
trainer = d2l.Trainer(max_epochs=100, gradient_clip_val=1, num_gpus=1)
trainer.fit(model, data)
```

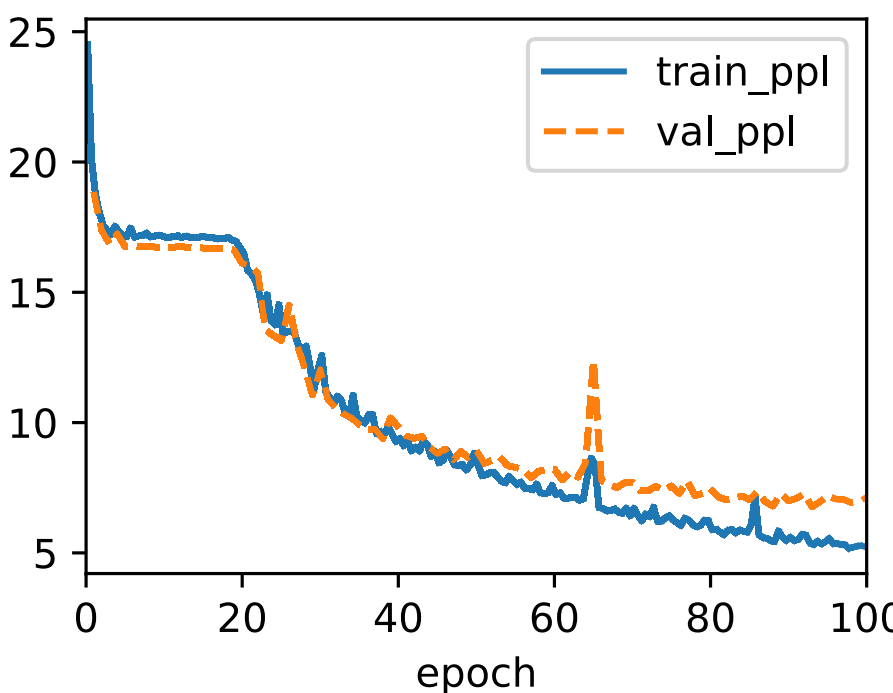

## 10.3.2 Concise Implementation

Fortunately many of the logistical details required to implement multiple layers of an RNN are readily available in high-level APIs. Our concise implementation will use such built-in functionalities. The code generalizes the one we used previously in Section 10.2, letting us specify the number of layers explicitly rather than picking the default of only one layer.

```
class GRU(d2l.RNN):  #@save
    """The multilayer GRU model."""
    def __init__(self, num_inputs, num_hiddens, num_layers, dropout=0):
        d2l.Module.__init__(self)
        self.save_hyperparameters()
        self.rnn = nn.GRU(num_inputs, num_hiddens, num_layers,
                          dropout=dropout)
```

The architectural decisions such as choosing hyperparameters are very similar to those of Section 10.2. We pick the same number of inputs and outputs as we have distinct tokens, i.e., `vocab_size`. The number of hidden units is still 32. The only difference is that we now select a nontrivial number of hidden layers by specifying the value of `num_layers`.

```
gru = GRU(num_inputs=len(data.vocab), num_hiddens=32, num_layers=2)
model = d2l.RNNLM(gru, vocab_size=len(data.vocab), lr=2)
trainer.fit(model, data)
```

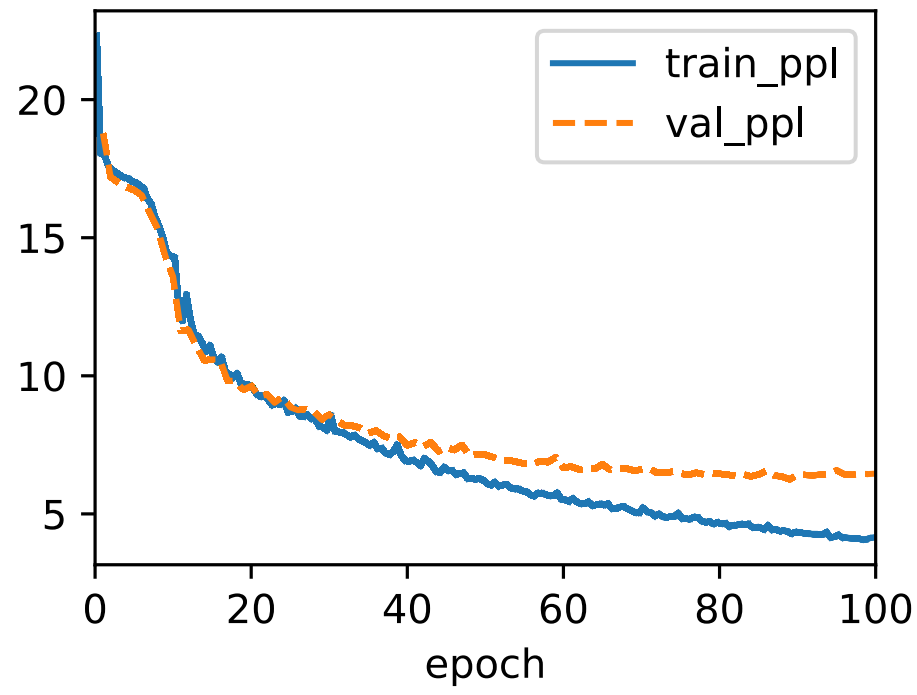

```
model.predict('it has', 20, data.vocab, d2l.try_gpu())
```

```
'it has some about the time'
```

## 10.3.3 Summary

In deep RNNs, the hidden state information is passed to the next time step of the current layer and the current time step of the next layer. There exist many different flavors of deep RNNs, such as LSTMs, GRUs, or vanilla RNNs. Conveniently, these models are all available as parts of the high-level APIs of deep learning frameworks. Initialization of models requires care. Overall, deep RNNs require considerable amount of work (such as learning rate and clipping) to ensure proper convergence.

## 10.3.4 Exercises

1. Replace the GRU by an LSTM and compare the accuracy and training speed.
2. Increase the training data to include multiple books. How low can you go on the perplexity scale?
3. Would you want to combine sources of different authors when modeling text? Why is this a good idea? What could go wrong?

Discussions[147].

# 10.4 Bidirectional Recurrent Neural Networks

So far, our working example of a sequence learning task has been language modeling, where we aim to predict the next token given all previous tokens in a sequence. In this scenario, we wish only to condition upon the leftward context, and thus the unidirectional chaining of a standard RNN seems appropriate. However, there are many other sequence learning tasks contexts where it is perfectly fine to condition the prediction at every time step on both the leftward and the rightward context. Consider, for example, part of speech detection. Why shouldn't we take the context in both directions into account when assessing the part of speech associated with a given word?

Another common task—often useful as a pretraining exercise prior to fine-tuning a model on an actual task of interest—is to mask out random tokens in a text document and then to train a sequence model to predict the values of the missing tokens. Note that depending on what comes after the blank, the likely value of the missing token changes dramatically:

- I am ___.
- I am ___ hungry.
- I am ___ hungry, and I can eat half a pig.

In the first sentence "happy" seems to be a likely candidate. The words "not" and "very" seem plausible in the second sentence, but "not" seems incompatible with the third sentences.

Fortunately, a simple technique transforms any unidirectional RNN into a bidirectional RNN (Schuster and Paliwal, 1997). We simply implement two unidirectional RNN layers chained together in opposite directions and acting on the same input (Fig. 10.4.1). For the first RNN layer, the first input is $\mathbf{x}_1$ and the last input is $\mathbf{x}_T$, but for the second RNN layer, the first input is $\mathbf{x}_T$ and the last input is $\mathbf{x}_1$. To produce the output of this bidirectional RNN layer, we simply concatenate together the corresponding outputs of the two underlying unidirectional RNN layers.

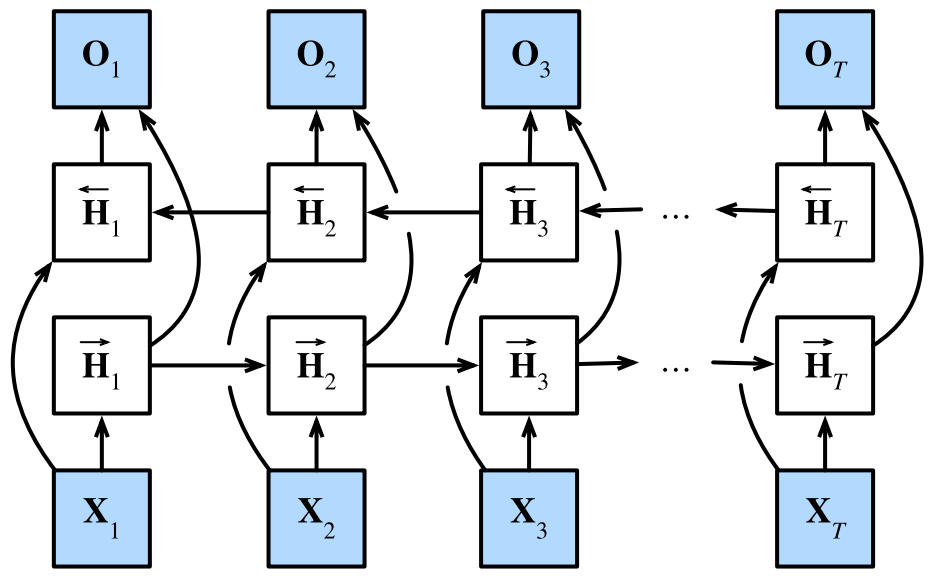

Fig. 10.4.1 Architecture of a bidirectional RNN.

Formally for any time step $t$, we consider a minibatch input $\mathbf{X}_t \in \mathbb{R}^{n\times d}$ (number of examples

$= n$; number of inputs in each example $= d$) and let the hidden layer activation function be $\phi$. In the bidirectional architecture, the forward and backward hidden states for this time step are $\overrightarrow{\mathbf{H}}_t \in \mathbb{R}^{n\times h}$ and $\overleftarrow{\mathbf{H}}_t \in \mathbb{R}^{n\times h}$, respectively, where $h$ is the number of hidden units. The forward and backward hidden state updates are as follows:

$$\begin{aligned}
\overrightarrow{\mathbf{H}}_t &= \phi(\mathbf{X}_t \mathbf{W}_{\textrm{xh}}^{(f)} + \overrightarrow{\mathbf{H}}_{t-1} \mathbf{W}_{\textrm{hh}}^{(f)} + \mathbf{b}_\textrm{h}^{(f)}),\\
\overleftarrow{\mathbf{H}}_t &= \phi(\mathbf{X}_t \mathbf{W}_{\textrm{xh}}^{(b)} + \overleftarrow{\mathbf{H}}_{t+1} \mathbf{W}_{\textrm{hh}}^{(b)} + \mathbf{b}_\textrm{h}^{(b)}),
\end{aligned} \tag{10.4.1}$$

where the weights $\mathbf{W}_{\textrm{xh}}^{(f)} \in \mathbb{R}^{d \times h}, \mathbf{W}_{\textrm{hh}}^{(f)} \in \mathbb{R}^{h \times h}, \mathbf{W}_{\textrm{xh}}^{(b)} \in \mathbb{R}^{d \times h}$, and $\mathbf{W}_{\textrm{hh}}^{(b)} \in \mathbb{R}^{h \times h}$, and the biases $\mathbf{b}_\textrm{h}^{(f)} \in \mathbb{R}^{1 \times h}$ and $\mathbf{b}_\textrm{h}^{(b)} \in \mathbb{R}^{1 \times h}$ are all the model parameters.

Next, we concatenate the forward and backward hidden states $\overrightarrow{\mathbf{H}}_t$ and $\overleftarrow{\mathbf{H}}_t$ to obtain the hidden state $\mathbf{H}_t \in \mathbb{R}^{n \times 2h}$ for feeding into the output layer. In deep bidirectional RNNs with multiple hidden layers, such information is passed on as *input* to the next bidirectional layer. Last, the output layer computes the output $\mathbf{O}_t \in \mathbb{R}^{n \times q}$ (number of outputs $= q$):

$$\mathbf{O}_t = \mathbf{H}_t \mathbf{W}_{\textrm{hq}} + \mathbf{b}_\textrm{q}. \tag{10.4.2}$$

Here, the weight matrix $\mathbf{W}_{\textrm{hq}} \in \mathbb{R}^{2h \times q}$ and the bias $\mathbf{b}_\textrm{q} \in \mathbb{R}^{1 \times q}$ are the model parameters of the output layer. While technically, the two directions can have different numbers of hidden units, this design choice is seldom made in practice. We now demonstrate a simple implementation of a bidirectional RNN.

```
import torch
from torch import nn
from d2l import torch as d2l
```

## 10.4.1 Implementation from Scratch

To implement a bidirectional RNN from scratch, we can include two unidirectional `RNNScratch` instances with separate learnable parameters.

```
class BiRNNScratch(d2l.Module):
    def __init__(self, num_inputs, num_hiddens, sigma=0.01):
        super().__init__()
        self.save_hyperparameters()
        self.f_rnn = d2l.RNNScratch(num_inputs, num_hiddens, sigma)
        self.b_rnn = d2l.RNNScratch(num_inputs, num_hiddens, sigma)
        self.num_hiddens *= 2  # The output dimension will be doubled
```

States of forward and backward RNNs are updated separately, while outputs of these two RNNs are concatenated.

```
@d2l.add_to_class(BiRNNScratch)
def forward(self, inputs, Hs=None):
    f_H, b_H = Hs if Hs is not None else (None, None)
```

```
    f_outputs, f_H = self.f_rnn(inputs, f_H)
    b_outputs, b_H = self.b_rnn(reversed(inputs), b_H)
    outputs = [torch.cat((f, b), -1) for f, b in zip(
        f_outputs, reversed(b_outputs))]
    return outputs, (f_H, b_H)
```

## 10.4.2 Concise Implementation

Using the high-level APIs, we can implement bidirectional RNNs more concisely. Here we take a GRU model as an example.

```
class BiGRU(d2l.RNN):
    def __init__(self, num_inputs, num_hiddens):
        d2l.Module.__init__(self)
        self.save_hyperparameters()
        self.rnn = nn.GRU(num_inputs, num_hiddens, bidirectional=True)
        self.num_hiddens *= 2
```

## 10.4.3 Summary

In bidirectional RNNs, the hidden state for each time step is simultaneously determined by the data prior to and after the current time step. Bidirectional RNNs are mostly useful for sequence encoding and the estimation of observations given bidirectional context. Bidirectional RNNs are very costly to train due to long gradient chains.

## 10.4.4 Exercises

1. If the different directions use a different number of hidden units, how will the shape of $\mathbf{H}_t$ change?
2. Design a bidirectional RNN with multiple hidden layers.
3. Polysemy is common in natural languages. For example, the word "bank" has different meanings in contexts "i went to the bank to deposit cash" and "i went to the bank to sit down". How can we design a neural network model such that given a context sequence and a word, a vector representation of the word in the correct context will be returned? What type of neural architectures is preferred for handling polysemy?

Discussions[148].

# 10.5 Machine Translation and the Dataset

Among the major breakthroughs that prompted widespread interest in modern RNNs was a major advance in the applied field of statistical *machine translation*. Here, the model is presented with a sentence in one language and must predict the corresponding sentence in another. Note that here the sentences may be of different lengths, and that corresponding words in the two sentences may not occur in the same order, owing to differences in the two language's grammatical structure.

Many problems have this flavor of mapping between two such "unaligned" sequences. Examples include mapping from dialog prompts to replies or from questions to answers. Broadly, such problems are called *sequence-to-sequence* (seq2seq) problems and they are our focus for both the remainder of this chapter and much of Chapter 11.

In this section, we introduce the machine translation problem and an example dataset that we will use in the subsequent examples. For decades, statistical formulations of translation between languages had been popular (Brown *et al.*, 1990, Brown *et al.*, 1988), even before researchers got neural network approaches working (methods were often lumped together under the term *neural machine translation*).

First we will need some new code to process our data. Unlike the language modeling that we saw in Section 9.3, here each example consists of two separate text sequences, one in the source language and another (the translation) in the target language. The following code snippets will show how to load the preprocessed data into minibatches for training.

```
import os
import torch
from d2l import torch as d2l
```

## 10.5.1 Downloading and Preprocessing the Dataset

To begin, we download an English–French dataset that consists of bilingual sentence pairs from the Tatoeba Project[149]. Each line in the dataset is a tab-delimited pair consisting of an English text sequence (the *source*) and the translated French text sequence (the *target*). Note that each text sequence can be just one sentence, or a paragraph of multiple sentences.

```
class MTFraEng(d2l.DataModule):  #@save
    """The English-French dataset."""
    def _download(self):
        d2l.extract(d2l.download(
            d2l.DATA_URL+'fra-eng.zip', self.root,
            '94646ad1522d915e7b0f9296181140edcf86a4f5'))
        with open(self.root + '/fra-eng/fra.txt', encoding='utf-8') as f:
            return f.read()
```

```
data = MTFraEng()
raw_text = data._download()
print(raw_text[:75])
```

```
Go.	Va !
Hi.	Salut !
Run!	Cours !
Run!	Courez !
Who?	Qui ?
Wow!	Ça alors !
```

After downloading the dataset, we proceed with several preprocessing steps for the raw text data. For instance, we replace non-breaking space with space, convert uppercase letters to lowercase ones, and insert space between words and punctuation marks.

```
@d2l.add_to_class(MTFraEng)  #@save
def _preprocess(self, text):
    # Replace non-breaking space with space
    text = text.replace('\u202f', ' ').replace('\xa0', ' ')
    # Insert space between words and punctuation marks
    no_space = lambda char, prev_char: char in ',.!?' and prev_char != ' '
    out = [' ' + char if i > 0 and no_space(char, text[i - 1]) else char
           for i, char in enumerate(text.lower())]
    return ''.join(out)
```

```
text = data._preprocess(raw_text)
print(text[:80])
```

```
go .	va !
hi .	salut !
run !	cours !
run !	courez !
who ?	qui ?
wow !	ça alors !
```

## 10.5.2 Tokenization

Unlike the character-level tokenization in Section 9.3, for machine translation we prefer word-level tokenization here (today's state-of-the-art models use more complex tokenization techniques). The following `_tokenize` method tokenizes the first `max_examples` text sequence pairs, where each token is either a word or a punctuation mark. We append the special "<eos>" token to the end of every sequence to indicate the end of the sequence. When a model is predicting by generating a sequence token after token, the generation of the "<eos>" token can suggest that the output sequence is complete. In the end, the method below returns two lists of token lists: `src` and `tgt`. Specifically, `src[i]` is a list of tokens from the $i^\text{th}$ text sequence in the source language (English here) and `tgt[i]` is that in the target language (French here).

```
@d2l.add_to_class(MTFraEng)  #@save
def _tokenize(self, text, max_examples=None):
    src, tgt = [], []
    for i, line in enumerate(text.split('\n')):
        if max_examples and i > max_examples: break
        parts = line.split('\t')
        if len(parts) == 2:
            # Skip empty tokens
            src.append([t for t in f'{parts[0]} <eos>'.split(' ') if t])
            tgt.append([t for t in f'{parts[1]} <eos>'.split(' ') if t])
    return src, tgt
```

```
src, tgt = data._tokenize(text)
src[:6], tgt[:6]
```

```
([['go', '.', '<eos>'],
  ['hi', '.', '<eos>'],
  ['run', '!', '<eos>'],
  ['run', '!', '<eos>'],
  ['who', '?', '<eos>'],
  ['wow', '!', '<eos>']],
 [['va', '!', '<eos>'],
  ['salut', '!', '<eos>'],
  ['cours', '!', '<eos>'],
  ['courez', '!', '<eos>'],
  ['qui', '?', '<eos>'],
  ['ça', 'alors', '!', '<eos>']])
```

Let's plot the histogram of the number of tokens per text sequence. In this simple English–French dataset, most of the text sequences have fewer than 20 tokens.

```
#@save
def show_list_len_pair_hist(legend, xlabel, ylabel, xlist, ylist):
    """Plot the histogram for list length pairs."""
    d2l.set_figsize()
    _, _, patches = d2l.plt.hist(
        [[len(l) for l in xlist], [len(l) for l in ylist]])
    d2l.plt.xlabel(xlabel)
    d2l.plt.ylabel(ylabel)
    for patch in patches[1].patches:
        patch.set_hatch('/')
    d2l.plt.legend(legend)
```

```
show_list_len_pair_hist(['source', 'target'], '# tokens per sequence',
                        'count', src, tgt);
```

### 10.5.3 Loading Sequences of Fixed Length

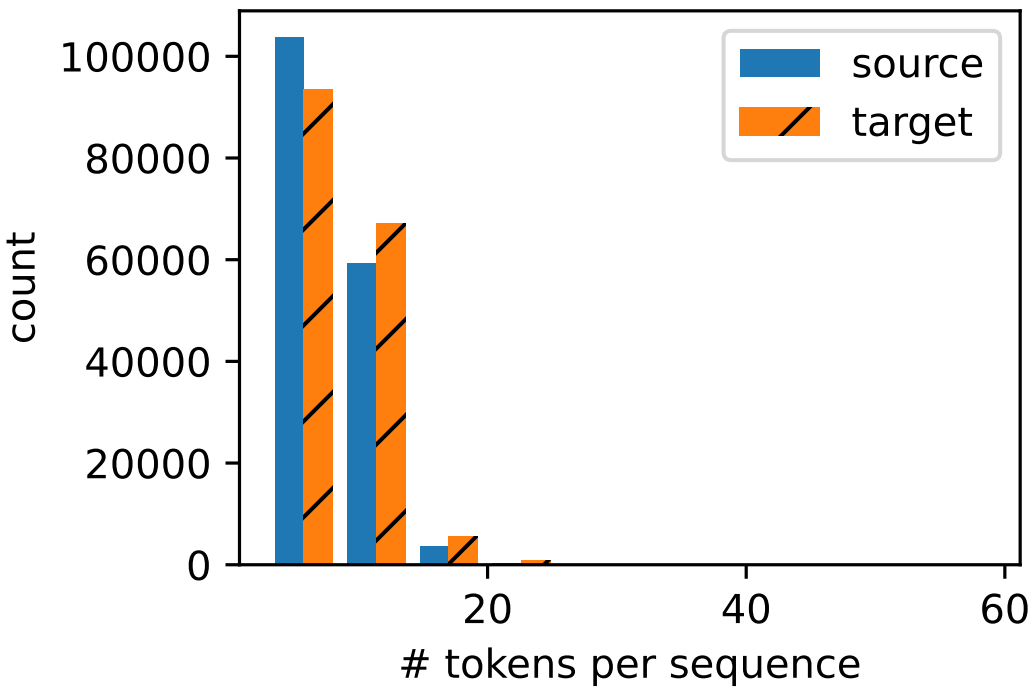

Recall that in language modeling each example sequence, either a segment of one sentence or a span over multiple sentences, had a fixed length. This was specified by the `num_steps` (number of time steps or tokens) argument from Section 9.3. In machine translation, each example is a pair of source and target text sequences, where the two text sequences may have different lengths.

For computational efficiency, we can still process a minibatch of text sequences at one time by *truncation* and *padding*. Suppose that every sequence in the same minibatch should have the same length `num_steps`. If a text sequence has fewer than `num_steps` tokens, we will keep appending the special "<pad>" token to its end until its length reaches `num_steps`. Otherwise, we will truncate the text sequence by only taking its first `num_steps` tokens and discarding the remaining. In this way, every text sequence will have the same length to be loaded in minibatches of the same shape. Furthermore, we also record length of the source sequence excluding padding tokens. This information will be needed by some models that we will cover later.

Since the machine translation dataset consists of pairs of languages, we can build two vocabularies for both the source language and the target language separately. With word-level tokenization, the vocabulary size will be significantly larger than that using character-level tokenization. To alleviate this, here we treat infrequent tokens that appear less than twice as the same unknown ("<unk>") token. As we will explain later (Fig. 10.7.1), when training with target sequences, the decoder output (label tokens) can be the same decoder input (target tokens), shifted by one token; and the special beginning-of-sequence "<bos>" token will be used as the first input token for predicting the target sequence (Fig. 10.7.3).

```
@d2l.add_to_class(MTFraEng)  #@save
def __init__(self, batch_size, num_steps=9, num_train=512, num_val=128):
    super(MTFraEng, self).__init__()
    self.save_hyperparameters()
    self.arrays, self.src_vocab, self.tgt_vocab = self._build_arrays(
        self._download())
```

```
@d2l.add_to_class(MTFraEng)  #@save
def _build_arrays(self, raw_text, src_vocab=None, tgt_vocab=None):
    def _build_array(sentences, vocab, is_tgt=False):
        pad_or_trim = lambda seq, t: (
            seq[:t] if len(seq) > t else seq + ['<pad>'] * (t - len(seq)))
        sentences = [pad_or_trim(s, self.num_steps) for s in sentences]
        if is_tgt:
            sentences = [['<bos>'] + s for s in sentences]
        if vocab is None:
            vocab = d2l.Vocab(sentences, min_freq=2)
        array = torch.tensor([vocab[s] for s in sentences])
        valid_len = (array != vocab['<pad>']).type(torch.int32).sum(1)
        return array, vocab, valid_len
    src, tgt = self._tokenize(self._preprocess(raw_text),
                              self.num_train + self.num_val)
    src_array, src_vocab, src_valid_len = _build_array(src, src_vocab)
    tgt_array, tgt_vocab, _ = _build_array(tgt, tgt_vocab, True)
    return ((src_array, tgt_array[:,:-1], src_valid_len, tgt_array[:,1:]),
            src_vocab, tgt_vocab)
```

### 10.5.4 Reading the Dataset

Finally, we define the `get_dataloader` method to return the data iterator.

```
@d2l.add_to_class(MTFraEng)  #@save
def get_dataloader(self, train):
    idx = slice(0, self.num_train) if train else slice(self.num_train, None)
    return self.get_tensorloader(self.arrays, train, idx)
```

Let's read the first minibatch from the English–French dataset.

```
data = MTFraEng(batch_size=3)
src, tgt, src_valid_len, label = next(iter(data.train_dataloader()))
print('source:', src.type(torch.int32))
print('decoder input:', tgt.type(torch.int32))
print('source len excluding pad:', src_valid_len.type(torch.int32))
print('label:', label.type(torch.int32))
```

```
source: tensor([[101, 175,   2,   3,   4,   4,   4,   4,   4],
        [ 86,  22,   2,   3,   4,   4,   4,   4,   4],
        [183, 105,   2,   3,   4,   4,   4,   4,   4]], dtype=torch.int32)
decoder input: tensor([[  3,   6,   0,   4,   5,   5,   5,   5,   5],
        [  3, 108, 183,  38,   2,   4,   5,   5,   5],
        [  3, 135,  92,  64,   2,   4,   5,   5,   5]], dtype=torch.int32)
source len excluding pad: tensor([4, 4, 4], dtype=torch.int32)
label: tensor([[  6,   0,   4,   5,   5,   5,   5,   5,   5],
        [108, 183,  38,   2,   4,   5,   5,   5,   5],
        [135,  92,  64,   2,   4,   5,   5,   5,   5]], dtype=torch.int32)
```

We show a pair of source and target sequences processed by the above `_build_arrays` method (in the string format).

```
@d2l.add_to_class(MTFraEng)  #@save
def build(self, src_sentences, tgt_sentences):
    raw_text = '\n'.join([src + '\t' + tgt for src, tgt in zip(
        src_sentences, tgt_sentences)])
    arrays, _, _ = self._build_arrays(
        raw_text, self.src_vocab, self.tgt_vocab)
    return arrays
```

```
src, tgt, _,  _ = data.build(['hi .'], ['salut .'])
print('source:', data.src_vocab.to_tokens(src[0].type(torch.int32)))
print('target:', data.tgt_vocab.to_tokens(tgt[0].type(torch.int32)))
```

```
source: ['hi', '.', '<eos>', '<pad>', '<pad>', '<pad>', '<pad>', '<pad>', '
↪<pad>']
target: ['<bos>', 'salut', '.', '<eos>', '<pad>', '<pad>', '<pad>', '<pad>', '
↪<pad>']
```

## 10.5.5 Summary

In natural language processing, *machine translation* refers to the task of automatically mapping from a sequence representing a string of text in a *source* language to a string representing a plausible translation in a *target* language. Using word-level tokenization, the vocabulary size will be significantly larger than that using character-level tokenization, but the sequence lengths will be much shorter. To mitigate the large vocabulary size, we can treat infrequent tokens as some "unknown" token. We can truncate and pad text sequences so that all of them will have the same length to be loaded in minibatches. Modern implementations often bucket sequences with similar lengths to avoid wasting excessive computation on padding.

## 10.5.6 Exercises

1. Try different values of the `max_examples` argument in the `_tokenize` method. How does this affect the vocabulary sizes of the source language and the target language?

2. Text in some languages such as Chinese and Japanese does not have word boundary indicators (e.g., space). Is word-level tokenization still a good idea for such cases? Why or why not?

Discussions[150].

# 10.6 The Encoder–Decoder Architecture

In general sequence-to-sequence problems like machine translation (Section 10.5), inputs and outputs are of varying lengths that are unaligned. The standard approach to handling this sort of data is to design an *encoder–decoder* architecture (Fig. 10.6.1) consisting of two major components: an *encoder* that takes a variable-length sequence as input, and a *decoder* that acts as a conditional language model, taking in the encoded input and the leftwards context of the target sequence and predicting the subsequent token in the target sequence.

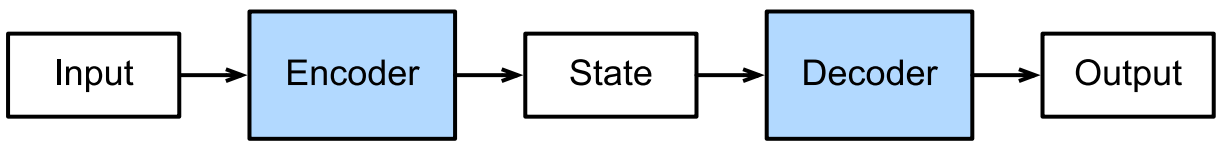

Fig. 10.6.1 The encoder–decoder architecture.

Let's take machine translation from English to French as an example. Given an input sequence in English: "They", "are", "watching", ".", this encoder–decoder architecture first encodes the variable-length input into a state, then decodes the state to generate the translated sequence, token by token, as output: "Ils", "regardent", ".". Since the encoder–decoder architecture forms the basis of different sequence-to-sequence models in subsequent sections, this section will convert this architecture into an interface that will be implemented later.

```
from torch import nn
from d2l import torch as d2l
```

## 10.6.1 Encoder

In the encoder interface, we just specify that the encoder takes variable-length sequences as input `X`. The implementation will be provided by any model that inherits this base `Encoder` class.

```
class Encoder(nn.Module):  #@save
    """The base encoder interface for the encoder--decoder architecture."""
    def __init__(self):
        super().__init__()

    # Later there can be additional arguments (e.g., length excluding padding)
    def forward(self, X, *args):
        raise NotImplementedError
```

## 10.6.2 Decoder

In the following decoder interface, we add an additional `init_state` method to convert the encoder output (`enc_all_outputs`) into the encoded state. Note that this step may require extra inputs, such as the valid length of the input, which was explained in Section 10.5. To generate a variable-length sequence token by token, every time the decoder may map an input (e.g., the generated token at the previous time step) and the encoded state into an output token at the current time step.

```
class Decoder(nn.Module):  #@save
    """The base decoder interface for the encoder--decoder architecture."""
    def __init__(self):
        super().__init__()

    # Later there can be additional arguments (e.g., length excluding padding)
    def init_state(self, enc_all_outputs, *args):
        raise NotImplementedError

    def forward(self, X, state):
        raise NotImplementedError
```

### 10.6.3 Putting the Encoder and Decoder Together

In the forward propagation, the output of the encoder is used to produce the encoded state, and this state will be further used by the decoder as one of its input.

```
class EncoderDecoder(d2l.Classifier):  #@save
    """The base class for the encoder--decoder architecture."""
    def __init__(self, encoder, decoder):
        super().__init__()
        self.encoder = encoder
        self.decoder = decoder

    def forward(self, enc_X, dec_X, *args):
        enc_all_outputs = self.encoder(enc_X, *args)
        dec_state = self.decoder.init_state(enc_all_outputs, *args)
        # Return decoder output only
        return self.decoder(dec_X, dec_state)[0]
```

In the next section, we will see how to apply RNNs to design sequence-to-sequence models based on this encoder–decoder architecture.

### 10.6.4 Summary

Encoder-decoder architectures can handle inputs and outputs that both consist of variable-length sequences and thus are suitable for sequence-to-sequence problems such as machine translation. The encoder takes a variable-length sequence as input and transforms it into a state with a fixed shape. The decoder maps the encoded state of a fixed shape to a variable-length sequence.

### 10.6.5 Exercises

1. Suppose that we use neural networks to implement the encoder–decoder architecture. Do the encoder and the decoder have to be the same type of neural network?
2. Besides machine translation, can you think of another application where the encoder–decoder architecture can be applied?

Discussions[151].

## 10.7 Sequence-to-Sequence Learning for Machine Translation

In so-called sequence-to-sequence problems such as machine translation (as discussed in Section 10.5), where inputs and outputs each consist of variable-length unaligned sequences, we generally rely on encoder–decoder architectures (Section 10.6). In this section, we will demonstrate the application of an encoder–decoder architecture, where both the encoder and decoder are implemented as RNNs, to the task of machine translation (Cho *et al.*, 2014, Sutskever *et al.*, 2014).

Here, the encoder RNN will take a variable-length sequence as input and transform it into a fixed-shape hidden state. Later, in Chapter 11, we will introduce attention mechanisms, which allow us to access encoded inputs without having to compress the entire input into a single fixed-length representation.

Then to generate the output sequence, one token at a time, the decoder model, consisting of a separate RNN, will predict each successive target token given both the input sequence and the preceding tokens in the output. During training, the decoder will typically be conditioned upon the preceding tokens in the official "ground truth" label. However, at test time, we will want to condition each output of the decoder on the tokens already predicted. Note that if we ignore the encoder, the decoder in a sequence-to-sequence architecture behaves just like a normal language model. Fig. 10.7.1 illustrates how to use two RNNs for sequence-to-sequence learning in machine translation.

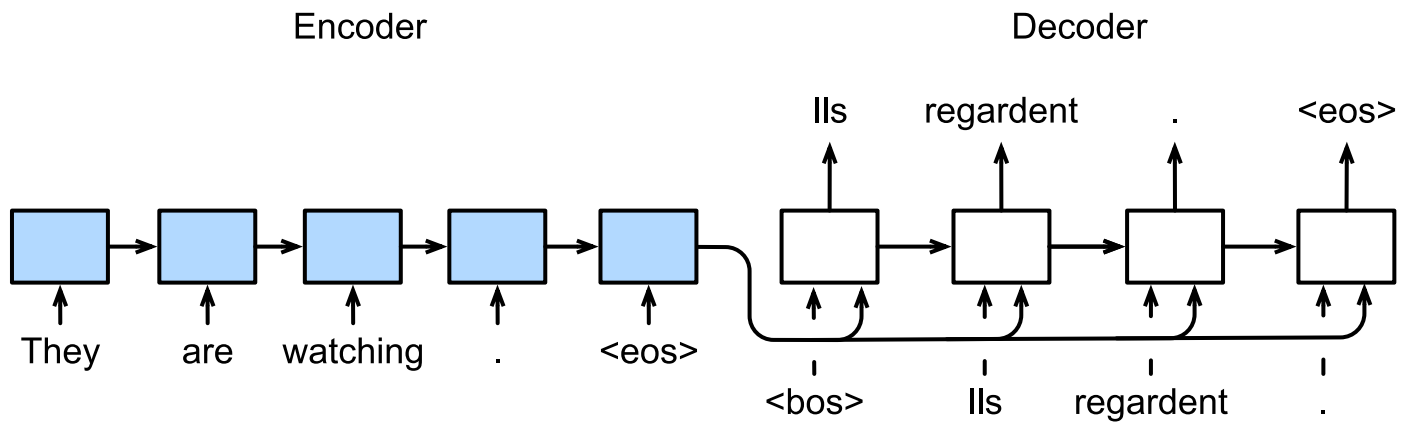

Fig. 10.7.1 Sequence-to-sequence learning with an RNN encoder and an RNN decoder.

In Fig. 10.7.1, the special "<eos>" token marks the end of the sequence. Our model can stop making predictions once this token is generated. At the initial time step of the RNN decoder, there are two special design decisions to be aware of: First, we begin every input with a special beginning-of-sequence “<bos>” token. Second, we may feed the final hidden state of the encoder into the decoder at every single decoding time step (Cho *et al.*, 2014). In some other designs, such as that of Sutskever *et al.* (2014), the final hidden state of the RNN encoder is used to initiate the hidden state of the decoder only at the first decoding step.

```
import collections
import math
import torch
from torch import nn
from torch.nn import functional as F
from d2l import torch as d2l
```

## 10.7.1 Teacher Forcing

While running the encoder on the input sequence is relatively straightforward, handling the input and output of the decoder requires more care. The most common approach is sometimes called *teacher forcing*. Here, the original target sequence (token labels) is fed into the decoder as input. More concretely, the special beginning-of-sequence token and the original target sequence, excluding the final token, are concatenated as input to the decoder, while the decoder output (labels for training) is the original target sequence, shifted by one token: “<bos>”, “Ils”, “regardent”, “.” → “Ils”, “regardent”, “.”, “<eos>” (Fig. 10.7.1).

Our implementation in Section 10.5.3 prepared training data for teacher forcing, where shifting tokens for self-supervised learning is similar to the training of language models in Section 9.3. An alternative approach is to feed the *predicted* token from the previous time step as the current input to the decoder.

In the following, we explain the design depicted in Fig. 10.7.1 in greater detail. We will train this model for machine translation on the English–French dataset as introduced in Section 10.5.

## 10.7.2 Encoder

Recall that the encoder transforms an input sequence of variable length into a fixed-shape *context variable* **c** (see Fig. 10.7.1).

Consider a single sequence example (batch size 1). Suppose the input sequence is $x_1, \ldots, x_T$, such that $x_t$ is the $t^{\textrm{th}}$ token. At time step $t$, the RNN transforms the input feature vector $\mathbf{x}_t$ for $x_t$ and the hidden state $\mathbf{h}_{t-1}$ from the previous time step into the current hidden state $\mathbf{h}_t$. We can use a function $f$ to express the transformation of the RNN's recurrent layer:

$$\mathbf{h}_t = f(\mathbf{x}_t, \mathbf{h}_{t-1}). \tag{10.7.1}$$

In general, the encoder transforms the hidden states at all time steps into a context variable through a customized function $q$:

$$\mathbf{c} = q(\mathbf{h}_1, \ldots, \mathbf{h}_T). \tag{10.7.2}$$

For example, in Fig. 10.7.1, the context variable is just the hidden state $\mathbf{h}_T$ corresponding to the encoder RNN's representation after processing the final token of the input sequence.

In this example, we have used a unidirectional RNN to design the encoder, where the hidden state only depends on the input subsequence at and before the time step of the hidden state. We can also construct encoders using bidirectional RNNs. In this case, a hidden state depends on the subsequence before and after the time step (including the input at the current time step), which encodes the information of the entire sequence.

Now let's implement the RNN encoder. Note that we use an *embedding layer* to obtain the feature vector for each token in the input sequence. The weight of an embedding layer is a matrix, where the number of rows corresponds to the size of the input vocabulary (`vocab_size`) and number of columns corresponds to the feature vector's dimension (`embed_size`). For any input token index $i$, the embedding layer fetches the $i^{\text{th}}$ row (starting from 0) of the weight matrix to return its feature vector. Here we implement the encoder with a multilayer GRU.

```
def init_seq2seq(module):  #@save
    """Initialize weights for sequence-to-sequence learning."""
    if type(module) == nn.Linear:
         nn.init.xavier_uniform_(module.weight)
    if type(module) == nn.GRU:
        for param in module._flat_weights_names:
            if "weight" in param:
                nn.init.xavier_uniform_(module._parameters[param])
```

```
class Seq2SeqEncoder(d2l.Encoder):  #@save
    """The RNN encoder for sequence-to-sequence learning."""
    def __init__(self, vocab_size, embed_size, num_hiddens, num_layers,
                 dropout=0):
        super().__init__()
        self.embedding = nn.Embedding(vocab_size, embed_size)
        self.rnn = d2l.GRU(embed_size, num_hiddens, num_layers, dropout)
        self.apply(init_seq2seq)

    def forward(self, X, *args):
        # X shape: (batch_size, num_steps)
        embs = self.embedding(X.t().type(torch.int64))
        # embs shape: (num_steps, batch_size, embed_size)
        outputs, state = self.rnn(embs)
        # outputs shape: (num_steps, batch_size, num_hiddens)
        # state shape: (num_layers, batch_size, num_hiddens)
        return outputs, state
```

Let's use a concrete example to illustrate the above encoder implementation. Below, we instantiate a two-layer GRU encoder whose number of hidden units is 16. Given a minibatch of sequence inputs X (batch size = 4; number of time steps = 9), the hidden states of the final layer at all the time steps (`enc_outputs` returned by the encoder's recurrent layers) are a tensor of shape (number of time steps, batch size, number of hidden units).

```
vocab_size, embed_size, num_hiddens, num_layers = 10, 8, 16, 2
batch_size, num_steps = 4, 9
encoder = Seq2SeqEncoder(vocab_size, embed_size, num_hiddens, num_layers)
X = torch.zeros((batch_size, num_steps))
enc_outputs, enc_state = encoder(X)
d2l.check_shape(enc_outputs, (num_steps, batch_size, num_hiddens))
```

Since we are using a GRU here, the shape of the multilayer hidden states at the final time step is (number of hidden layers, batch size, number of hidden units).

```
d2l.check_shape(enc_state, (num_layers, batch_size, num_hiddens))
```

## 10.7.3 Decoder

Given a target output sequence $y_1, y_2, \ldots, y_{T'}$ for each time step $t'$ (we use $t'$ to differentiate from the input sequence time steps), the decoder assigns a predicted probability to each possible token occurring at step $y_{t'+1}$ conditioned upon the previous tokens in the target $y_1, \ldots, y_{t'}$ and the context variable $\mathbf{c}$, i.e., $P(y_{t'+1} \mid y_1, \ldots, y_{t'}, \mathbf{c})$.

To predict the subsequent token $t'+1$ in the target sequence, the RNN decoder takes the previous step's target token $y_{t'}$, the hidden RNN state from the previous time step $\mathbf{s}_{t'-1}$, and the context variable $\mathbf{c}$ as its input, and transforms them into the hidden state $\mathbf{s}_{t'}$ at the current time step. We can use a function $g$ to express the transformation of the decoder's hidden layer:

$$\mathbf{s}_{t'} = g(y_{t'-1}, \mathbf{c}, \mathbf{s}_{t'-1}). \tag{10.7.3}$$

After obtaining the hidden state of the decoder, we can use an output layer and the softmax operation to compute the predictive distribution $p(y_{t'+1} \mid y_1, \ldots, y_{t'}, \mathbf{c})$ over the subsequent output token $t'+1$.

Following Fig. 10.7.1, when implementing the decoder as follows, we directly use the hidden state at the final time step of the encoder to initialize the hidden state of the decoder. This requires that the RNN encoder and the RNN decoder have the same number of layers and hidden units. To further incorporate the encoded input sequence information, the context variable is concatenated with the decoder input at all the time steps. To predict the probability distribution of the output token, we use a fully connected layer to transform the hidden state at the final layer of the RNN decoder.

```
class Seq2SeqDecoder(d2l.Decoder):
    """The RNN decoder for sequence to sequence learning."""
    def __init__(self, vocab_size, embed_size, num_hiddens, num_layers,
```

```
                 dropout=0):
        super().__init__()
        self.embedding = nn.Embedding(vocab_size, embed_size)
        self.rnn = d2l.GRU(embed_size+num_hiddens, num_hiddens,
                           num_layers, dropout)
        self.dense = nn.LazyLinear(vocab_size)
        self.apply(init_seq2seq)

    def init_state(self, enc_all_outputs, *args):
        return enc_all_outputs

    def forward(self, X, state):
        # X shape: (batch_size, num_steps)
        # embs shape: (num_steps, batch_size, embed_size)
        embs = self.embedding(X.t().type(torch.int32))
        enc_output, hidden_state = state
        # context shape: (batch_size, num_hiddens)
        context = enc_output[-1]
        # Broadcast context to (num_steps, batch_size, num_hiddens)
        context = context.repeat(embs.shape[0], 1, 1)
        # Concat at the feature dimension
        embs_and_context = torch.cat((embs, context), -1)
        outputs, hidden_state = self.rnn(embs_and_context, hidden_state)
        outputs = self.dense(outputs).swapaxes(0, 1)
        # outputs shape: (batch_size, num_steps, vocab_size)
        # hidden_state shape: (num_layers, batch_size, num_hiddens)
        return outputs, [enc_output, hidden_state]
```

To illustrate the implemented decoder, below we instantiate it with the same hyperparameters from the aforementioned encoder. As we can see, the output shape of the decoder becomes (batch size, number of time steps, vocabulary size), where the final dimension of the tensor stores the predicted token distribution.

```
decoder = Seq2SeqDecoder(vocab_size, embed_size, num_hiddens, num_layers)
state = decoder.init_state(encoder(X))
dec_outputs, state = decoder(X, state)
d2l.check_shape(dec_outputs, (batch_size, num_steps, vocab_size))
d2l.check_shape(state[1], (num_layers, batch_size, num_hiddens))
```

The layers in the above RNN encoder–decoder model are summarized in Fig. 10.7.2.

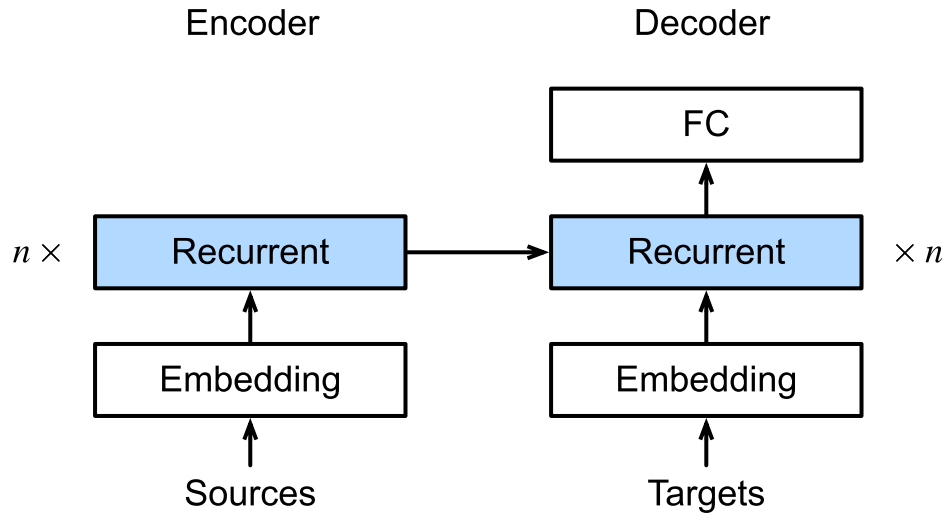

Fig. 10.7.2 Layers in an RNN encoder–decoder model.

### 10.7.4 Encoder–Decoder for Sequence-to-Sequence Learning

Putting it all together in code yields the following:

```
class Seq2Seq(d2l.EncoderDecoder):  #@save
    """The RNN encoder--decoder for sequence to sequence learning."""
    def __init__(self, encoder, decoder, tgt_pad, lr):
        super().__init__(encoder, decoder)
        self.save_hyperparameters()

    def validation_step(self, batch):
        Y_hat = self(*batch[:-1])
        self.plot('loss', self.loss(Y_hat, batch[-1]), train=False)

    def configure_optimizers(self):
        # Adam optimizer is used here
        return torch.optim.Adam(self.parameters(), lr=self.lr)
```

### 10.7.5 Loss Function with Masking

At each time step, the decoder predicts a probability distribution for the output tokens. As with language modeling, we can apply softmax to obtain the distribution and calculate the cross-entropy loss for optimization. Recall from Section 10.5 that the special padding tokens are appended to the end of sequences and so sequences of varying lengths can be efficiently loaded in minibatches of the same shape. However, prediction of padding tokens should be excluded from loss calculations. To this end, we can mask irrelevant entries with zero values so that multiplication of any irrelevant prediction with zero equates to zero.

```
@d2l.add_to_class(Seq2Seq)
def loss(self, Y_hat, Y):
    l = super(Seq2Seq, self).loss(Y_hat, Y, averaged=False)
    mask = (Y.reshape(-1) != self.tgt_pad).type(torch.float32)
    return (l * mask).sum() / mask.sum()
```

### 10.7.6 Training

Now we can create and train an RNN encoder–decoder model for sequence-to-sequence learning on the machine translation dataset.

```
data = d2l.MTFraEng(batch_size=128)
embed_size, num_hiddens, num_layers, dropout = 256, 256, 2, 0.2
encoder = Seq2SeqEncoder(
    len(data.src_vocab), embed_size, num_hiddens, num_layers, dropout)
decoder = Seq2SeqDecoder(
    len(data.tgt_vocab), embed_size, num_hiddens, num_layers, dropout)
model = Seq2Seq(encoder, decoder, tgt_pad=data.tgt_vocab['<pad>'],
                lr=0.005)
```

```
trainer = d2l.Trainer(max_epochs=30, gradient_clip_val=1, num_gpus=1)
trainer.fit(model, data)
```

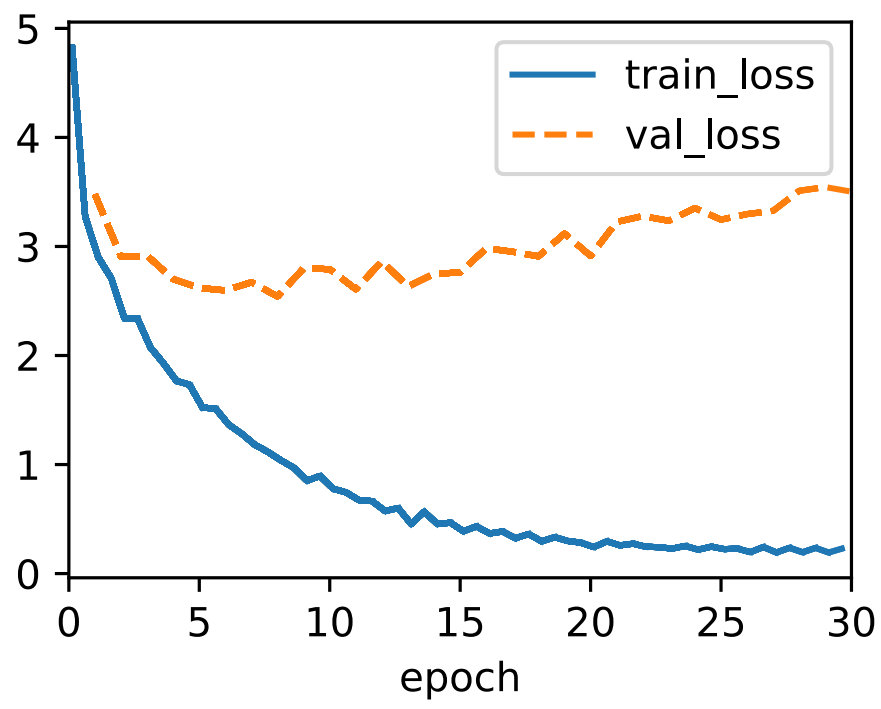

## 10.7.7 Prediction

To predict the output sequence at each step, the predicted token from the previous time step is fed into the decoder as an input. One simple strategy is to sample whichever token that has been assigned by the decoder the highest probability when predicting at each step. As in training, at the initial time step the beginning-of-sequence ("<bos>") token is fed into the decoder. This prediction process is illustrated in Fig. 10.7.3. When the end-of-sequence ("<eos>") token is predicted, the prediction of the output sequence is complete.

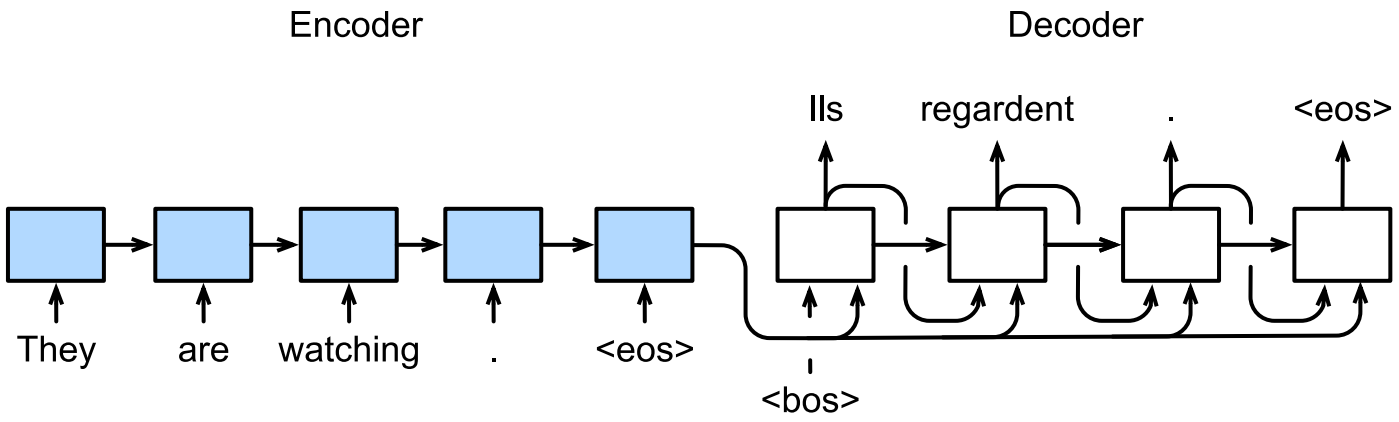

Fig. 10.7.3 Predicting the output sequence token by token using an RNN encoder–decoder.

In the next section, we will introduce more sophisticated strategies based on beam search (Section 10.8).

```
@d2l.add_to_class(d2l.EncoderDecoder)  #@save
def predict_step(self, batch, device, num_steps,
                 save_attention_weights=False):
    batch = [a.to(device) for a in batch]
    src, tgt, src_valid_len, _ = batch
    enc_all_outputs = self.encoder(src, src_valid_len)
    dec_state = self.decoder.init_state(enc_all_outputs, src_valid_len)
```

```
    outputs, attention_weights = [tgt[:, 0].unsqueeze(1), ], []
    for _ in range(num_steps):
        Y, dec_state = self.decoder(outputs[-1], dec_state)
        outputs.append(Y.argmax(2))
        # Save attention weights (to be covered later)
        if save_attention_weights:
            attention_weights.append(self.decoder.attention_weights)
    return torch.cat(outputs[1:], 1), attention_weights
```

### 10.7.8 Evaluation of Predicted Sequences

We can evaluate a predicted sequence by comparing it with the target sequence (the ground truth). But what precisely is the appropriate measure for comparing similarity between two sequences?

Bilingual Evaluation Understudy (BLEU), though originally proposed for evaluating machine translation results (Papineni *et al.*, 2002), has been extensively used in measuring the quality of output sequences for different applications. In principle, for any $n$-gram (Section 9.3.1) in the predicted sequence, BLEU evaluates whether this $n$-gram appears in the target sequence.

Denote by $p_n$ the precision of an $n$-gram, defined as the ratio of the number of matched $n$-grams in the predicted and target sequences to the number of $n$-grams in the predicted sequence. To explain, given a target sequence $A$, $B$, $C$, $D$, $E$, $F$, and a predicted sequence $A$, $B$, $B$, $C$, $D$, we have $p_1 = 4/5$, $p_2 = 3/4$, $p_3 = 1/3$, and $p_4 = 0$. Now let $\textrm{len}_{\textrm{label}}$ and $\textrm{len}_{\textrm{pred}}$ be the numbers of tokens in the target sequence and the predicted sequence, respectively. Then, BLEU is defined as

$$\exp\left(\min\left(0, 1 - \frac{\textrm{len}_{\textrm{label}}}{\textrm{len}_{\textrm{pred}}}\right)\right) \prod_{n=1}^k p_n^{1/2^n}, \tag{10.7.4}$$

where $k$ is the longest $n$-gram for matching.

Based on the definition of BLEU in (10.7.4), whenever the predicted sequence is the same as the target sequence, BLEU is 1. Moreover, since matching longer $n$-grams is more difficult, BLEU assigns a greater weight when a longer $n$-gram has high precision. Specifically, when $p_n$ is fixed, $p_n^{1/2^n}$ increases as $n$ grows (the original paper uses $p_n^{1/n}$). Furthermore, since predicting shorter sequences tends to yield a higher $p_n$ value, the coefficient before the multiplication term in (10.7.4) penalizes shorter predicted sequences. For example, when $k = 2$, given the target sequence $A$, $B$, $C$, $D$, $E$, $F$ and the predicted sequence $A$, $B$, although $p_1 = p_2 = 1$, the penalty factor $\exp(1 - 6/2) \approx 0.14$ lowers the BLEU.

We implement the BLEU measure as follows.

```
def bleu(pred_seq, label_seq, k):  #@save
    """Compute the BLEU."""
```

```
    pred_tokens, label_tokens = pred_seq.split(' '), label_seq.split(' ')
    len_pred, len_label = len(pred_tokens), len(label_tokens)
    score = math.exp(min(0, 1 - len_label / len_pred))
    for n in range(1, min(k, len_pred) + 1):
        num_matches, label_subs = 0, collections.defaultdict(int)
        for i in range(len_label - n + 1):
            label_subs[' '.join(label_tokens[i: i + n])] += 1
        for i in range(len_pred - n + 1):
            if label_subs[' '.join(pred_tokens[i: i + n])] > 0:
                num_matches += 1
                label_subs[' '.join(pred_tokens[i: i + n])] -= 1
        score *= math.pow(num_matches / (len_pred - n + 1), math.pow(0.5, n))
    return score
```

In the end, we use the trained RNN encoder–decoder to translate a few English sentences into French and compute the BLEU of the results.

```
engs = ['go .', 'i lost .', 'he\'s calm .', 'i\'m home .']
fras = ['va !', 'j\'ai perdu .', 'il est calme .', 'je suis chez moi .']
preds, _ = model.predict_step(
    data.build(engs, fras), d2l.try_gpu(), data.num_steps)
for en, fr, p in zip(engs, fras, preds):
    translation = []
    for token in data.tgt_vocab.to_tokens(p):
        if token == '<eos>':
            break
        translation.append(token)
    print(f'{en} => {translation}, bleu,'
          f'{bleu(" ".join(translation), fr, k=2):.3f}')
```

```
go . => ['va', '!'], bleu,1.000
i lost . => ["j'ai", 'perdu', '.'], bleu,1.000
he's calm . => ['soyez', 'calmes', '.'], bleu,0.000
i'm home . => ['je', 'suis', 'chez', 'moi', '.'], bleu,1.000
```

### 10.7.9 Summary

Following the design of the encoder–decoder architecture, we can use two RNNs to design a model for sequence-to-sequence learning. In encoder–decoder training, the teacher forcing approach feeds original output sequences (in contrast to predictions) into the decoder. When implementing the encoder and the decoder, we can use multilayer RNNs. We can use masks to filter out irrelevant computations, such as when calculating the loss. For evaluating output sequences, BLEU is a popular measure that matches $n$-grams between the predicted sequence and the target sequence.

### 10.7.10 Exercises

1. Can you adjust the hyperparameters to improve the translation results?
2. Rerun the experiment without using masks in the loss calculation. What results do you observe? Why?
3. If the encoder and the decoder differ in the number of layers or the number of hidden units, how can we initialize the hidden state of the decoder?
4. In training, replace teacher forcing with feeding the prediction at the previous time step into the decoder. How does this influence the performance?
5. Rerun the experiment by replacing GRU with LSTM.
6. Are there any other ways to design the output layer of the decoder?

Discussions[152].

# 10.8 Beam Search

In Section 10.7, we introduced the encoder–decoder architecture, and the standard techniques for training them end-to-end. However, when it came to test-time prediction, we mentioned only the *greedy* strategy, where we select at each time step the token given the highest predicted probability of coming next, until, at some time step, we find that we have predicted the special end-of-sequence “<eos>” token. In this section, we will begin by formalizing this *greedy search* strategy and identifying some problems that practitioners tend to run into. Subsequently, we compare this strategy with two alternatives: *exhaustive search* (illustrative but not practical) and *beam search* (the standard method in practice).

Let’s begin by setting up our mathematical notation, borrowing conventions from Section 10.7. At any time step $t'$, the decoder outputs predictions representing the probability of each token in the vocabulary coming next in the sequence (the likely value of $y_{t'+1}$), conditioned on the previous tokens $y_1, \ldots, y_{t'}$ and the context variable $\mathbf{c}$, produced by the encoder to represent the input sequence. To quantify computational cost, denote by $\mathcal{Y}$ the output vocabulary (including the special end-of-sequence token “<eos>”). Let’s also specify the maximum number of tokens of an output sequence as $T'$. Our goal is to search for an ideal output from all $O(|\mathcal{Y}|^{T'})$ possible output sequences. Note that this slightly overestimates the number of distinct outputs because there are no subsequent tokens once the “<eos>” token occurs. However, for our purposes, this number roughly captures the size of the search space.

## 10.8.1 Greedy Search

Consider the simple *greedy search* strategy from Section 10.7. Here, at any time step $t'$, we simply select the token with the highest conditional probability from $\mathcal{Y}$, i.e.,

$$y_{t'} = \operatorname*{argmax}_{y \in \mathcal{Y}} P(y \mid y_1, \ldots, y_{t'-1}, \mathbf{c}). \tag{10.8.1}$$

Once our model outputs "<eos>" (or we reach the maximum length $T'$) the output sequence is completed.

This strategy might look reasonable, and in fact it is not so bad! Considering how computationally undemanding it is, you'd be hard pressed to get more bang for your buck. However, if we put aside efficiency for a minute, it might seem more reasonable to search for the *most likely sequence*, not the sequence of (greedily selected) *most likely tokens*. It turns out that these two objects can be quite different. The most likely sequence is the one that maximizes the expression $\prod_{t'=1}^{T'} P(y_{t'} \mid y_1, \ldots, y_{t'-1}, \mathbf{c})$. In our machine translation example, if the decoder truly recovered the probabilities of the underlying generative process, then this would give us the most likely translation. Unfortunately, there is no guarantee that greedy search will give us this sequence.

Let's illustrate it with an example. Suppose that there are four tokens "A", "B", "C", and "<eos>" in the output dictionary. In Fig. 10.8.1, the four numbers under each time step represent the conditional probabilities of generating "A", "B", "C", and "<eos>" respectively, at that time step.

| Time step | 1 | 2 | 3 | 4 |
|---|---|---|---|---|
| A | 0.5 | 0.1 | 0.2 | 0.0 |
| B | 0.2 | 0.4 | 0.2 | 0.2 |
| C | 0.2 | 0.3 | 0.4 | 0.2 |
| <eos> | 0.1 | 0.2 | 0.2 | 0.6 |

Fig. 10.8.1 At each time step, greedy search selects the token with the highest conditional probability.

At each time step, greedy search selects the token with the highest conditional probability. Therefore, the output sequence "A", "B", "C", and "<eos>" will be predicted (Fig. 10.8.1). The conditional probability of this output sequence is $0.5 \times 0.4 \times 0.4 \times 0.6 = 0.048$.

Next, let's look at another example in Fig. 10.8.2. Unlike in Fig. 10.8.1, at time step 2 we select the token "C", which has the *second* highest conditional probability.

| Time step | 1 | 2 | 3 | 4 |
|---|---|---|---|---|
| A | 0.5 | 0.1 | 0.1 | 0.1 |
| B | 0.2 | 0.4 | 0.6 | 0.2 |
| C | 0.2 | 0.3 | 0.2 | 0.1 |
| <eos> | 0.1 | 0.2 | 0.1 | 0.6 |

Fig. 10.8.2 The four numbers under each time step represent the conditional probabilities of generating "A", "B", "C", and "<eos>" at that time step. At time step 2, the token "C", which has the second highest conditional probability, is selected.

Since the output subsequences at time steps 1 and 2, on which time step 3 is based, have changed from "A" and "B" in Fig. 10.8.1 to "A" and "C" in Fig. 10.8.2, the conditional probability of each token at time step 3 has also changed in Fig. 10.8.2. Suppose that we choose the token "B" at time step 3. Now time step 4 is conditional on the output subsequence at the first three time steps "A", "C", and "B", which has changed from "A", "B", and "C" in Fig. 10.8.1. Therefore, the conditional probability of generating each token at time step 4 in Fig. 10.8.2 is also different from that in Fig. 10.8.1. As a result, the conditional probability of the output sequence "A", "C", "B", and "<eos>" in Fig. 10.8.2 is $0.5 \times 0.3 \times 0.6 \times 0.6 = 0.054$, which is greater than that of greedy search in Fig. 10.8.1. In this example, the output sequence "A", "B", "C", and "<eos>" obtained by the greedy search is not optimal.

## 10.8.2 Exhaustive Search

If the goal is to obtain the most likely sequence, we may consider using *exhaustive search*: enumerate all the possible output sequences with their conditional probabilities, and then output the one that scores the highest predicted probability.

While this would certainly give us what we desire, it would come at a prohibitive computational cost of $O(|\mathcal{Y}|^{T'})$, exponential in the sequence length and with an enormous base given by the vocabulary size. For example, when $|\mathcal{Y}| = 10000$ and $T' = 10$, both small numbers when compared with ones in real applications, we will need to evaluate $10000^{10} = 10^{40}$ sequences, which is already beyond the capabilities of any foreseeable computers. On the other hand, the computational cost of greedy search is $O(|\mathcal{Y}|\,T')$: miraculously cheap but far from optimal. For example, when $|\mathcal{Y}| = 10000$ and $T' = 10$, we only need to evaluate $10000 \times 10 = 10^5$ sequences.

## 10.8.3 Beam Search

You could view sequence decoding strategies as lying on a spectrum, with *beam search* striking a compromise between the efficiency of greedy search and the optimality of exhaustive search. The most straightforward version of beam search is characterized by a single hyperparameter, the *beam size*, $k$. Let's explain this terminology. At time step 1, we select the $k$ tokens with the highest predicted probabilities. Each of them will be the first token of $k$ candidate output sequences, respectively. At each subsequent time step, based on the $k$ candidate output sequences at the previous time step, we continue to select $k$ candidate output sequences with the highest predicted probabilities from $k\,|\mathcal{Y}|$ possible choices.

Fig. 10.8.3 demonstrates the process of beam search with an example. Suppose that the output vocabulary contains only five elements: $\mathcal{Y} = \{A, B, C, D, E\}$, where one of them is "<eos>". Let the beam size be two and the maximum length of an output sequence be three. At time step 1, suppose that the tokens with the highest conditional probabilities $P(y_1 \mid \mathbf{c})$ are $A$ and $C$. At time step 2, for all $y_2 \in \mathcal{Y}$, we compute

$$
\begin{aligned}
P(A, y_2 \mid \mathbf{c}) = P(A \mid \mathbf{c})P(y_2 \mid A, \mathbf{c}),\\
P(C, y_2 \mid \mathbf{c}) = P(C \mid \mathbf{c})P(y_2 \mid C, \mathbf{c}),
\end{aligned}
\tag{10.8.2}
$$

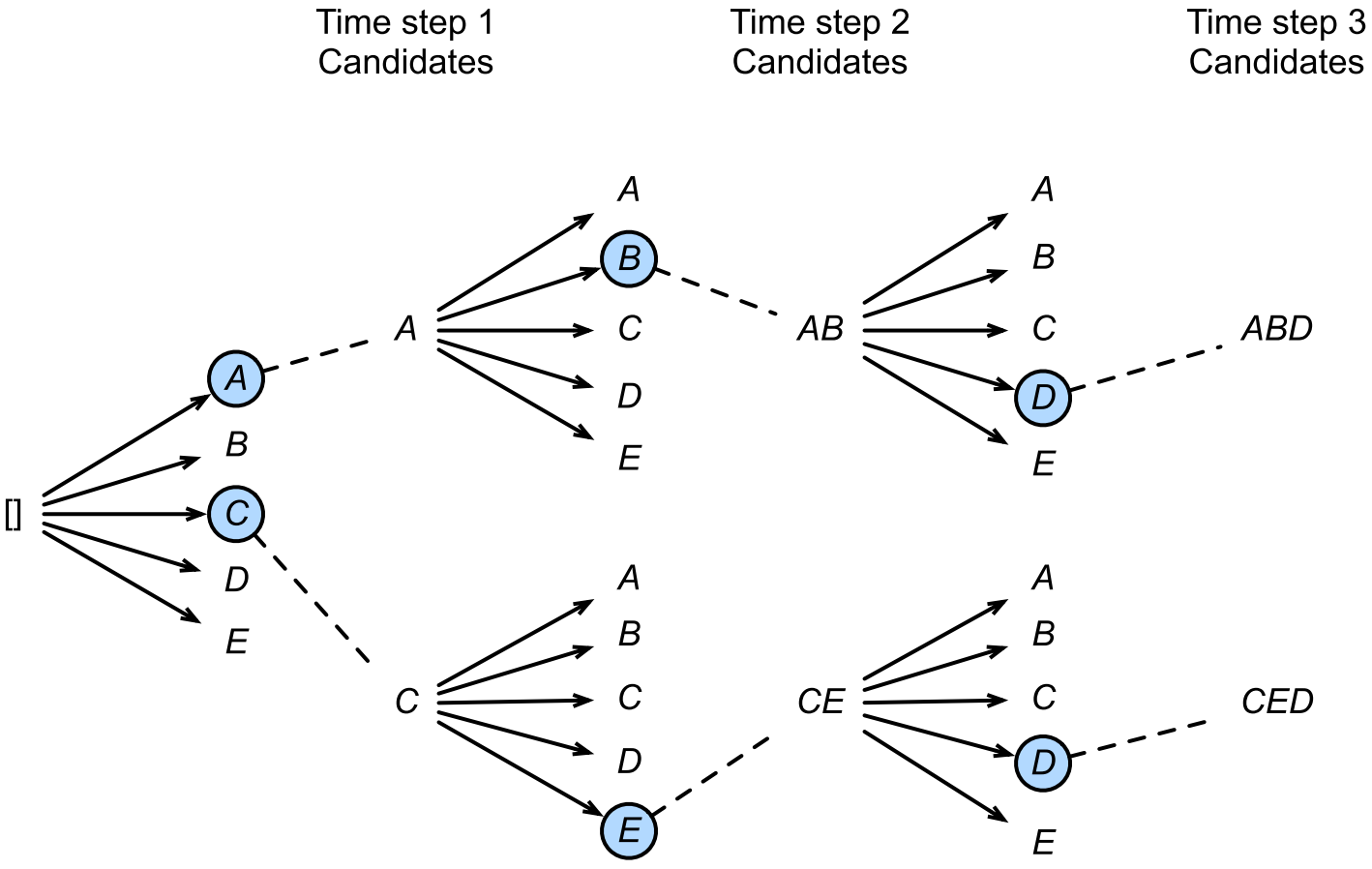

Fig. 10.8.3 The process of beam search (beam size = 2; maximum length of an output sequence = 3). The candidate output sequences are $A$, $C$, $AB$, $CE$, $ABD$, and $CED$.

and pick the largest two among these ten values, say $P(A, B \mid \mathbf{c})$ and $P(C, E \mid \mathbf{c})$. Then at time step 3, for all $y_3 \in \mathcal{Y}$, we compute

$$\begin{aligned} P(A, B, y_3 \mid \mathbf{c}) &= P(A, B \mid \mathbf{c})P(y_3 \mid A, B, \mathbf{c}), \\ P(C, E, y_3 \mid \mathbf{c}) &= P(C, E \mid \mathbf{c})P(y_3 \mid C, E, \mathbf{c}), \end{aligned} \tag{10.8.3}$$

and pick the largest two among these ten values, say $P(A, B, D \mid \mathbf{c})$ and $P(C, E, D \mid \mathbf{c})$. As a result, we get six candidates output sequences: (i) $A$; (ii) $C$; (iii) $A$, $B$; (iv) $C$, $E$; (v) $A$, $B$, $D$; and (vi) $C$, $E$, $D$.

In the end, we obtain the set of final candidate output sequences based on these six sequences (e.g., discard portions including and after "<eos>"). Then we choose the output sequence which maximizes the following score:

$$\frac{1}{L^\alpha} \log P(y_1, \ldots, y_L \mid \mathbf{c}) = \frac{1}{L^\alpha} \sum_{t'=1}^{L} \log P(y_{t'} \mid y_1, \ldots, y_{t'-1}, \mathbf{c}); \tag{10.8.4}$$

here $L$ is the length of the final candidate sequence and $\alpha$ is usually set to 0.75. Since a longer sequence has more logarithmic terms in the summation of (10.8.4), the term $L^\alpha$ in the denominator penalizes long sequences.

The computational cost of beam search is $O(k\,|\mathcal{Y}|\,T')$. This result is in between that of greedy search and that of exhaustive search. Greedy search can be treated as a special case of beam search arising when the beam size is set to 1.

## 10.8.4 Summary

Sequence searching strategies include greedy search, exhaustive search, and beam search. Beam search provides a trade-off between accuracy and computational cost via the flexible choice of the beam size.

### 10.8.5 Exercises

1. Can we treat exhaustive search as a special type of beam search? Why or why not?
2. Apply beam search in the machine translation problem in Section 10.7. How does the beam size affect the translation results and the prediction speed?
3. We used language modeling for generating text following user-provided prefixes in Section 9.5. Which kind of search strategy does it use? Can you improve it?

Discussions[153].

# 11 Attention Mechanisms and Transformers

The earliest years of the deep learning boom were driven primarily by results produced using the multilayer perceptron, convolutional network, and recurrent network architectures. Remarkably, the model architectures that underpinned many of deep learning's breakthroughs in the 2010s had changed remarkably little relative to their antecedents despite the lapse of nearly 30 years. While plenty of new methodological innovations made their way into most practitioner's toolkits—ReLU activations, residual layers, batch normalization, dropout, and adaptive learning rate schedules come to mind—the core underlying architectures were clearly recognizable as scaled-up implementations of classic ideas. Despite thousands of papers proposing alternative ideas, models resembling classical convolutional neural networks (Chapter 7) retained *state-of-the-art* status in computer vision and models resembling Sepp Hochreiter's original design for the LSTM recurrent neural network (Section 10.1), dominated most applications in natural language processing. Arguably, to that point, the rapid emergence of deep learning appeared to be primarily attributable to shifts in the available computational resources (thanks to innovations in parallel computing with GPUs) and the availability of massive data resources (thanks to cheap storage and Internet services). While these factors may indeed remain the primary drivers behind this technology's increasing power we are also witnessing, at long last, a sea change in the landscape of dominant architectures.

At the present moment, the dominant models for nearly all natural language processing tasks are based on the Transformer architecture. Given any new task in natural language processing, the default first-pass approach is to grab a large Transformer-based pretrained model, (e.g., BERT (Devlin *et al.*, 2018), ELECTRA (Clark *et al.*, 2020), RoBERTa (Liu *et al.*, 2019), or Longformer (Beltagy *et al.*, 2020)) adapting the output layers as necessary, and fine-tuning the model on the available data for the downstream task. If you have been paying attention to the last few years of breathless news coverage centered on OpenAI's large language models, then you have been tracking a conversation centered on the GPT-2 and GPT-3 Transformer-based models (Brown *et al.*, 2020, Radford *et al.*, 2019). Meanwhile, the vision Transformer has emerged as a default model for diverse vision tasks, including image recognition, object detection, semantic segmentation, and superresolution (Dosovitskiy *et al.*, 2021, Liu *et al.*, 2021). Transformers also showed up as competitive methods for speech recognition (Gulati *et al.*, 2020), reinforcement learning (Chen *et al.*, 2021), and graph neural networks (Dwivedi and Bresson, 2020).

The core idea behind the Transformer model is the *attention mechanism*, an innovation that was originally envisioned as an enhancement for encoder–decoder RNNs applied to sequence-to-sequence applications, such as machine translations (Bahdanau *et al.*, 2014). You might recall that in the first sequence-to-sequence models for machine translation (Sutskever *et al.*, 2014), the entire input was compressed by the encoder into a single fixed-length vector to be fed into the decoder. The intuition behind attention is that rather than compressing the input, it might be better for the decoder to revisit the input sequence at every step. Moreover, rather than always seeing the same representation of the input, one might imagine that the decoder should selectively focus on particular parts of the input sequence at particular decoding steps. Bahdanau's attention mechanism provided a simple means by which the decoder could dynamically *attend* to different parts of the input at each decoding step. The high-level idea is that the encoder could produce a representation of length equal to the original input sequence. Then, at decoding time, the decoder can (via some control mechanism) receive as input a context vector consisting of a weighted sum of the representations on the input at each time step. Intuitively, the weights determine the extent to which each step's context "focuses" on each input token, and the key is to make this process for assigning the weights differentiable so that it can be learned along with all of the other neural network parameters.

Initially, the idea was a remarkably successful enhancement to the recurrent neural networks that already dominated machine translation applications. The models performed better than the original encoder–decoder sequence-to-sequence architectures. Furthermore, researchers noted that some nice qualitative insights sometimes emerged from inspecting the pattern of attention weights. In translation tasks, attention models often assigned high attention weights to cross-lingual synonyms when generating the corresponding words in the target language. For example, when translating the sentence "my feet hurt" to "j'ai mal au pieds", the neural network might assign high attention weights to the representation of "feet" when generating the corresponding French word "pieds". These insights spurred claims that attention models confer "interpretability" although what precisely the attention weights mean—i.e., how, if at all, they should be *interpreted* remains a hazy research topic.

However, attention mechanisms soon emerged as more significant concerns, beyond their usefulness as an enhancement for encoder–decoder recurrent neural networks and their putative usefulness for picking out salient inputs. Vaswani *et al.* (2017) proposed the Transformer architecture for machine translation, dispensing with recurrent connections altogether, and instead relying on cleverly arranged attention mechanisms to capture all relationships among input and output tokens. The architecture performed remarkably well, and by 2018 the Transformer began showing up in the majority of state-of-the-art natural language processing systems. Moreover, at the same time, the dominant practice in natural language processing became to pretrain large-scale models on enormous generic background corpora to optimize some self-supervised pretraining objective, and then to fine-tune these models using the available downstream data. The gap between Transformers and traditional architectures grew especially wide when applied in this pretraining paradigm, and thus the ascendance of Transformers coincided with the ascendence of such large-scale pretrained models, now sometimes called *foundation models* (Bommasani *et al.*, 2021).

In this chapter, we introduce attention models, starting with the most basic intuitions and the simplest instantiations of the idea. We then work our way up to the Transformer archi-tecture, the vision Transformer, and the landscape of modern Transformer-based pretrained models.

## 11.1 Queries, Keys, and Values

So far all the networks we have reviewed crucially relied on the input being of a well-defined size. For instance, the images in ImageNet are of size $224 \times 224$ pixels and CNNs are specifically tuned to this size. Even in natural language processing the input size for RNNs is well defined and fixed. Variable size is addressed by sequentially processing one token at a time, or by specially designed convolution kernels (Kalchbrenner *et al.*, 2014). This approach can lead to significant problems when the input is truly of varying size with varying information content, such as in Section 10.7 in the transformation of text (Sutskever *et al.*, 2014). In particular, for long sequences it becomes quite difficult to keep track of everything that has already been generated or even viewed by the network. Even explicit tracking heuristics such as proposed by Yang *et al.* (2016) only offer limited benefit.

Compare this to databases. In their simplest form they are collections of keys ($k$) and values ($v$). For instance, our database $\mathcal{D}$ might consist of tuples $\{$("Zhang", "Aston"), ("Lipton", "Zachary"), ("Li", "Mu"), ("Smola", "Alex"), ("Hu", "Rachel"), ("Werness", "Brent")$\}$ with the last name being the key and the first name being the value. We can operate on $\mathcal{D}$, for instance with the exact query ($q$) for "Li" which would return the value "Mu". If ("Li", "Mu") was not a record in $\mathcal{D}$, there would be no valid answer. If we also allowed for approximate matches, we would retrieve ("Lipton", "Zachary") instead. This quite simple and trivial example nonetheless teaches us a number of useful things:

- We can design queries $q$ that operate on ($k$,$v$) pairs in such a manner as to be valid regardless of the database size.
- The same query can receive different answers, according to the contents of the database.
- The "code" being executed for operating on a large state space (the database) can be quite simple (e.g., exact match, approximate match, top-$k$).
- There is no need to compress or simplify the database to make the operations effective.

Clearly we would not have introduced a simple database here if it wasn't for the purpose of explaining deep learning. Indeed, this leads to one of the most exciting concepts introduced in deep learning in the past decade: the *attention mechanism* (Bahdanau *et al.*, 2014). We will cover the specifics of its application to machine translation later. For now, simply consider the following: denote by $\mathcal{D} \stackrel{\textrm{def}}{=} \{(\mathbf{k}_1, \mathbf{v}_1), \ldots (\mathbf{k}_m, \mathbf{v}_m)\}$ a database of $m$ tuples of *keys* and *values*. Moreover, denote by $\mathbf{q}$ a *query*. Then we can define the *attention* over $\mathcal{D}$ as

$$\textrm{Attention}(\mathbf{q}, \mathcal{D}) \stackrel{\textrm{def}}{=} \sum_{i=1}^m \alpha(\mathbf{q}, \mathbf{k}_i) \mathbf{v}_i, \tag{11.1.1}$$

where $\alpha(\mathbf{q}, \mathbf{k}_i) \in \mathbb{R}$ ($i = 1, \ldots, m$) are scalar attention weights. The operation itself is typically referred to as *attention pooling*. The name *attention* derives from the fact that the operation pays particular attention to the terms for which the weight $\alpha$ is significant (i.e., large). As such, the attention over $\mathcal{D}$ generates a linear combination of values contained in the database. In fact, this contains the above example as a special case where all but one weight is zero. We have a number of special cases:

- The weights $\alpha(\mathbf{q}, \mathbf{k}_i)$ are nonnegative. In this case the output of the attention mechanism is contained in the convex cone spanned by the values $\mathbf{v}_i$.
- The weights $\alpha(\mathbf{q}, \mathbf{k}_i)$ form a convex combination, i.e., $\sum_i \alpha(\mathbf{q}, \mathbf{k}_i) = 1$ and $\alpha(\mathbf{q}, \mathbf{k}_i) \geq 0$ for all $i$. This is the most common setting in deep learning.
- Exactly one of the weights $\alpha(\mathbf{q}, \mathbf{k}_i)$ is $1$, while all others are $0$. This is akin to a traditional database query.
- All weights are equal, i.e., $\alpha(\mathbf{q}, \mathbf{k}_i) = \frac{1}{m}$ for all $i$. This amounts to averaging across the entire database, also called average pooling in deep learning.

A common strategy for ensuring that the weights sum up to $1$ is to normalize them via

$$\alpha(\mathbf{q}, \mathbf{k}_i) = \frac{\alpha(\mathbf{q}, \mathbf{k}_i)}{\sum_j \alpha(\mathbf{q}, \mathbf{k}_j)}. \tag{11.1.2}$$

In particular, to ensure that the weights are also nonnegative, one can resort to exponentiation. This means that we can now pick *any* function $a(\mathbf{q}, \mathbf{k})$ and then apply the softmax operation used for multinomial models to it via

$$\alpha(\mathbf{q}, \mathbf{k}_i) = \frac{\exp(a(\mathbf{q}, \mathbf{k}_i))}{\sum_j \exp(a(\mathbf{q}, \mathbf{k}_j))}. \tag{11.1.3}$$

This operation is readily available in all deep learning frameworks. It is differentiable and its gradient never vanishes, all of which are desirable properties in a model. Note though, the attention mechanism introduced above is not the only option. For instance, we can design a non-differentiable attention model that can be trained using reinforcement learning methods (Mnih *et al.*, 2014). As one would expect, training such a model is quite complex. Consequently the bulk of modern attention research follows the framework outlined in Fig. 11.1.1. We thus focus our exposition on this family of differentiable mechanisms.

What is quite remarkable is that the actual "code" for executing on the set of keys and values, namely the query, can be quite concise, even though the space to operate on is significant. This is a desirable property for a network layer as it does not require too many parameters to learn. Just as convenient is the fact that attention can operate on arbitrarily large databases without the need to change the way the attention pooling operation is performed.

```
import torch
from d2l import torch as d2l
```

## 11.1.1 Visualization

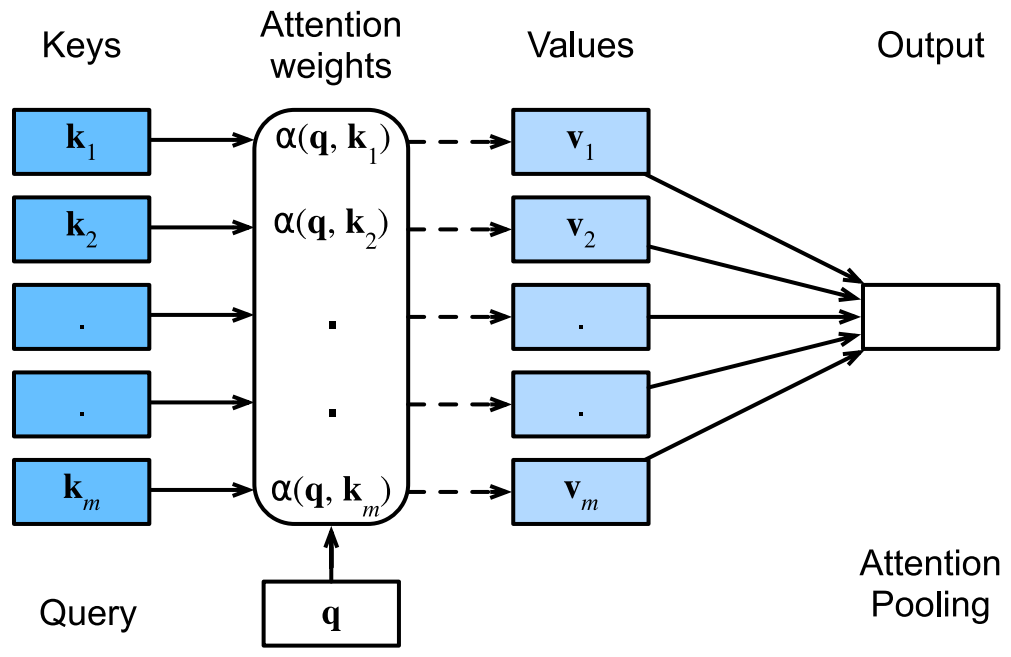

Fig. 11.1.1 The attention mechanism computes a linear combination over values $\mathbf{v}_i$ via attention pooling, where weights are derived according to the compatibility between a query $\mathbf{q}$ and keys $\mathbf{k}_i$.

---

One of the benefits of the attention mechanism is that it can be quite intuitive, particularly when the weights are nonnegative and sum to $1$. In this case we might *interpret* large weights as a way for the model to select components of relevance. While this is a good intuition, it is important to remember that it is just that, an *intuition*. Regardless, we may want to visualize its effect on the given set of keys when applying a variety of different queries. This function will come in handy later.

We thus define the `show_heatmaps` function. Note that it does not take a matrix (of attention weights) as its input but rather a tensor with four axes, allowing for an array of different queries and weights. Consequently the input `matrices` has the shape (number of rows for display, number of columns for display, number of queries, number of keys). This will come in handy later on when we want to visualize the workings that are to design Transformers.

```
#@save
def show_heatmaps(matrices, xlabel, ylabel, titles=None, figsize=(2.5, 2.5),
                  cmap='Reds'):
    """Show heatmaps of matrices."""
    d2l.use_svg_display()
    num_rows, num_cols, _, _ = matrices.shape
    fig, axes = d2l.plt.subplots(num_rows, num_cols, figsize=figsize,
                                 sharex=True, sharey=True, squeeze=False)
    for i, (row_axes, row_matrices) in enumerate(zip(axes, matrices)):
        for j, (ax, matrix) in enumerate(zip(row_axes, row_matrices)):
            pcm = ax.imshow(matrix.detach().numpy(), cmap=cmap)
            if i == num_rows - 1:
                ax.set_xlabel(xlabel)
            if j == 0:
                ax.set_ylabel(ylabel)
            if titles:
                ax.set_title(titles[j])
    fig.colorbar(pcm, ax=axes, shrink=0.6);
```

As a quick sanity check let's visualize the identity matrix, representing a case where the attention weight is $1$ only when the query and the key are the same.

```
attention_weights = torch.eye(10).reshape((1, 1, 10, 10))
show_heatmaps(attention_weights, xlabel='Keys', ylabel='Queries')
```

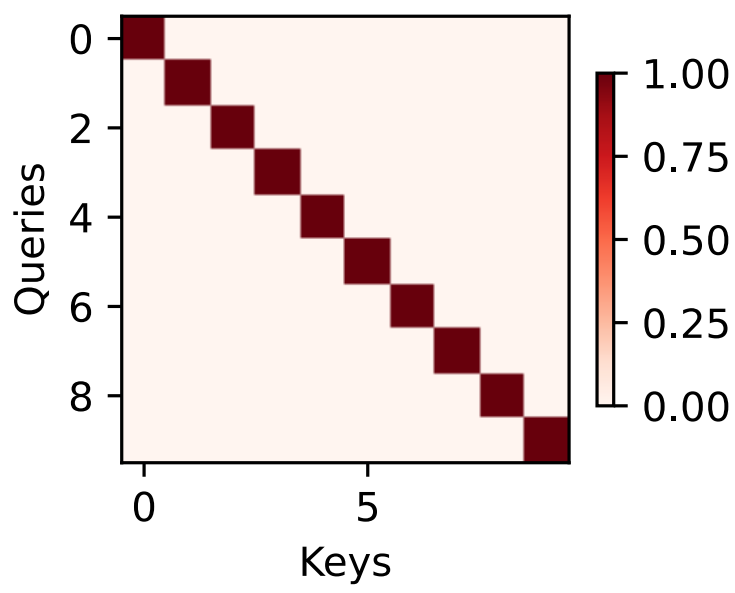

## 11.1.2 Summary

The attention mechanism allows us to aggregate data from many (key, value) pairs. So far our discussion was quite abstract, simply describing a way to pool data. We have not explained yet where those mysterious queries, keys, and values might arise from. Some intuition might help here: for instance, in a regression setting, the query might correspond to the location where the regression should be carried out. The keys are the locations where past data was observed and the values are the (regression) values themselves. This is the so-called Nadaraya–Watson estimator (Nadaraya, 1964, Watson, 1964) that we will be studying in the next section.

By design, the attention mechanism provides a *differentiable* means of control by which a neural network can select elements from a set and to construct an associated weighted sum over representations.

## 11.1.3 Exercises

1. Suppose that you wanted to reimplement approximate (key, query) matches as used in classical databases, which attention function would you pick?
2. Suppose that the attention function is given by $a(\mathbf{q}, \mathbf{k}_i) = \mathbf{q}^\top \mathbf{k}_i$ and that $\mathbf{k}_i = \mathbf{v}_i$ for $i = 1, \ldots, m$. Denote by $p(\mathbf{k}_i; \mathbf{q})$ the probability distribution over keys when using the softmax normalization in (11.1.3). Prove that $\nabla_{\mathbf{q}} \mathrm{Attention}(\mathbf{q}, \mathcal{D}) = \mathrm{Cov}_{p(\mathbf{k}_i; \mathbf{q})}[\mathbf{k}_i]$.
3. Design a differentiable search engine using the attention mechanism.
4. Review the design of the Squeeze and Excitation Networks (Hu *et al.*, 2018) and interpret them through the lens of the attention mechanism.

Discussions[154].

# 11.2 Attention Pooling by Similarity

Now that we have introduced the primary components of the attention mechanism, let's use them in a rather classical setting, namely regression and classification via kernel density estimation (Nadaraya, 1964, Watson, 1964). This detour simply provides additional background: it is entirely optional and can be skipped if needed. At their core, Nadaraya–Watson estimators rely on some similarity kernel $\alpha(\mathbf{q}, \mathbf{k})$ relating queries $\mathbf{q}$ to keys $\mathbf{k}$. Some common kernels are

$$\begin{aligned}
\alpha(\mathbf{q}, \mathbf{k}) & = \exp\left(-\frac{1}{2} \|\mathbf{q} - \mathbf{k}\|^2 \right) && \textrm{Gaussian;} \\
\alpha(\mathbf{q}, \mathbf{k}) & = 1 \textrm{ if } \|\mathbf{q} - \mathbf{k}\| \leq 1 && \textrm{Boxcar;} \\
\alpha(\mathbf{q}, \mathbf{k}) & = \mathop{\mathrm{max}}\left(0, 1 - \|\mathbf{q} - \mathbf{k}\|\right) && \textrm{Epanechikov.}
\end{aligned} \tag{11.2.1}$$

There are many more choices that we could pick. See a Wikipedia article[155] for a more extensive review and how the choice of kernels is related to kernel density estimation, sometimes also called *Parzen Windows* (Parzen, 1957). All of the kernels are heuristic and can be tuned. For instance, we can adjust the width, not only on a global basis but even on a per-coordinate basis. Regardless, all of them lead to the following equation for regression and classification alike:

$$f(\mathbf{q}) = \sum_i \mathbf{v}_i \frac{\alpha(\mathbf{q}, \mathbf{k}_i)}{\sum_j \alpha(\mathbf{q}, \mathbf{k}_j)}. \tag{11.2.2}$$

In the case of a (scalar) regression with observations $(\mathbf{x}_i, y_i)$ for features and labels respectively, $\mathbf{v}_i = y_i$ are scalars, $\mathbf{k}_i = \mathbf{x}_i$ are vectors, and the query $\mathbf{q}$ denotes the new location where $f$ should be evaluated. In the case of (multiclass) classification, we use one-hot-encoding of $y_i$ to obtain $\mathbf{v}_i$. One of the convenient properties of this estimator is that it requires no training. Even more so, if we suitably narrow the kernel with increasing amounts of data, the approach is consistent (Mack and Silverman, 1982), i.e., it will converge to some statistically optimal solution. Let's start by inspecting some kernels.

```python
import numpy as np
import torch
from torch import nn
from torch.nn import functional as F
from d2l import torch as d2l

d2l.use_svg_display()
```

## 11.2.1 Kernels and Data

All the kernels $\alpha(\mathbf{k}, \mathbf{q})$ defined in this section are *translation and rotation invariant*; that is, if we shift and rotate $\mathbf{k}$ and $\mathbf{q}$ in the same manner, the value of $\alpha$ remains unchanged. For simplicity we thus pick scalar arguments $k, q \in \mathbb{R}$ and pick the key $k = 0$ as the origin. This yields:

```
# Define some kernels
def gaussian(x):
    return torch.exp(-x**2 / 2)

def boxcar(x):
    return torch.abs(x) < 1.0

def constant(x):
    return 1.0 + 0 * x

def epanechikov(x):
    return torch.max(1 - torch.abs(x), torch.zeros_like(x))
```

```
fig, axes = d2l.plt.subplots(1, 4, sharey=True, figsize=(12, 3))

kernels = (gaussian, boxcar, constant, epanechikov)
names = ('Gaussian', 'Boxcar', 'Constant', 'Epanechikov')
x = torch.arange(-2.5, 2.5, 0.1)
for kernel, name, ax in zip(kernels, names, axes):
    ax.plot(x.detach().numpy(), kernel(x).detach().numpy())
    ax.set_xlabel(name)

d2l.plt.show()
```

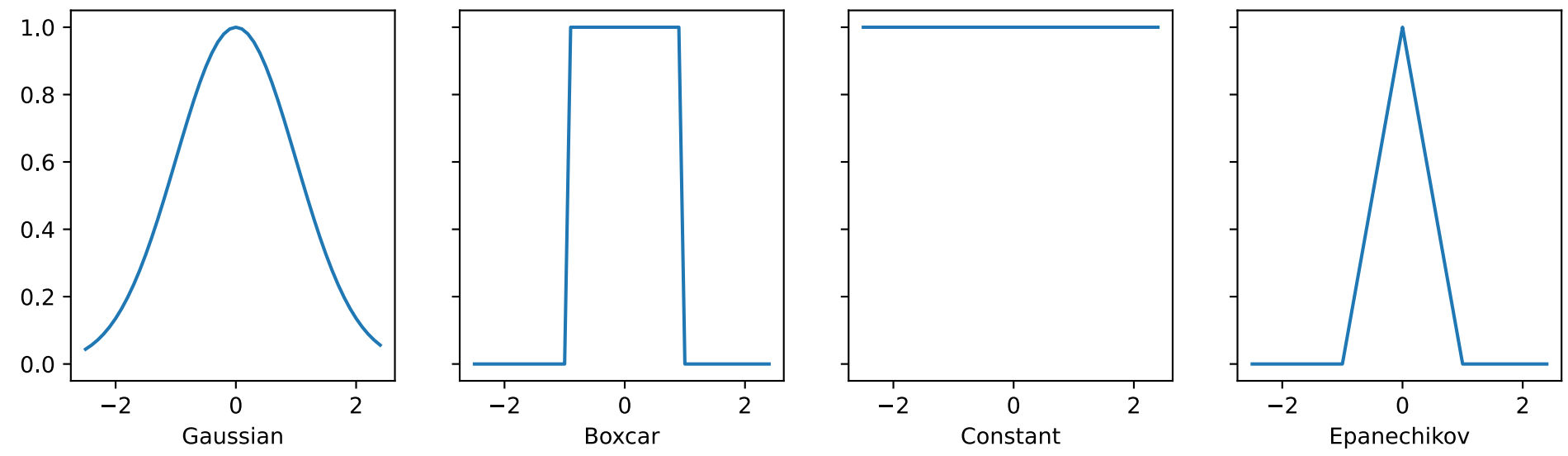

Different kernels correspond to different notions of range and smoothness. For instance, the boxcar kernel only attends to observations within a distance of 1 (or some otherwise defined hyperparameter) and does so indiscriminately.

To see Nadaraya–Watson estimation in action, let's define some training data. In the following we use the dependency

$$y_i = 2\sin(x_i) + x_i + \epsilon, \tag{11.2.3}$$

where $\epsilon$ is drawn from a normal distribution with zero mean and unit variance. We draw 40 training examples.

```
def f(x):
    return 2 * torch.sin(x) + x

n = 40
x_train, _ = torch.sort(torch.rand(n) * 5)
y_train = f(x_train) + torch.randn(n)
x_val = torch.arange(0, 5, 0.1)
y_val = f(x_val)
```

### 11.2.2 Attention Pooling via Nadaraya–Watson Regression

Now that we have data and kernels, all we need is a function that computes the kernel regression estimates. Note that we also want to obtain the relative kernel weights in order to perform some minor diagnostics. Hence we first compute the kernel between all training features (covariates) `x_train` and all validation features `x_val`. This yields a matrix, which we subsequently normalize. When multiplied with the training labels `y_train` we obtain the estimates.

Recall attention pooling in (11.1.1). Let each validation feature be a query, and each training feature–label pair be a key–value pair. As a result, the normalized relative kernel weights (`attention_w` below) are the *attention weights*.

```
def nadaraya_watson(x_train, y_train, x_val, kernel):
    dists = x_train.reshape((-1, 1)) - x_val.reshape((1, -1))
    # Each column/row corresponds to each query/key
    k = kernel(dists).type(torch.float32)
    # Normalization over keys for each query
    attention_w = k / k.sum(0)
    y_hat = y_train@attention_w
    return y_hat, attention_w
```

Let's have a look at the kind of estimates that the different kernels produce.

```
def plot(x_train, y_train, x_val, y_val, kernels, names, attention=False):
    fig, axes = d2l.plt.subplots(1, 4, sharey=True, figsize=(12, 3))
    for kernel, name, ax in zip(kernels, names, axes):
        y_hat, attention_w = nadaraya_watson(x_train, y_train, x_val, kernel)
        if attention:
            pcm = ax.imshow(attention_w.detach().numpy(), cmap='Reds')
        else:
            ax.plot(x_val, y_hat)
            ax.plot(x_val, y_val, 'm--')
            ax.plot(x_train, y_train, 'o', alpha=0.5);
        ax.set_xlabel(name)
        if not attention:
            ax.legend(['y_hat', 'y'])
    if attention:
        fig.colorbar(pcm, ax=axes, shrink=0.7)
```

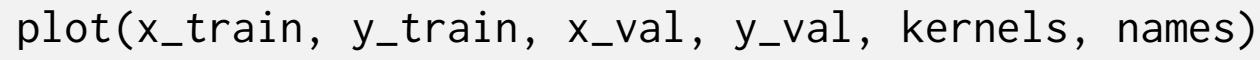

```
plot(x_train, y_train, x_val, y_val, kernels, names)
```

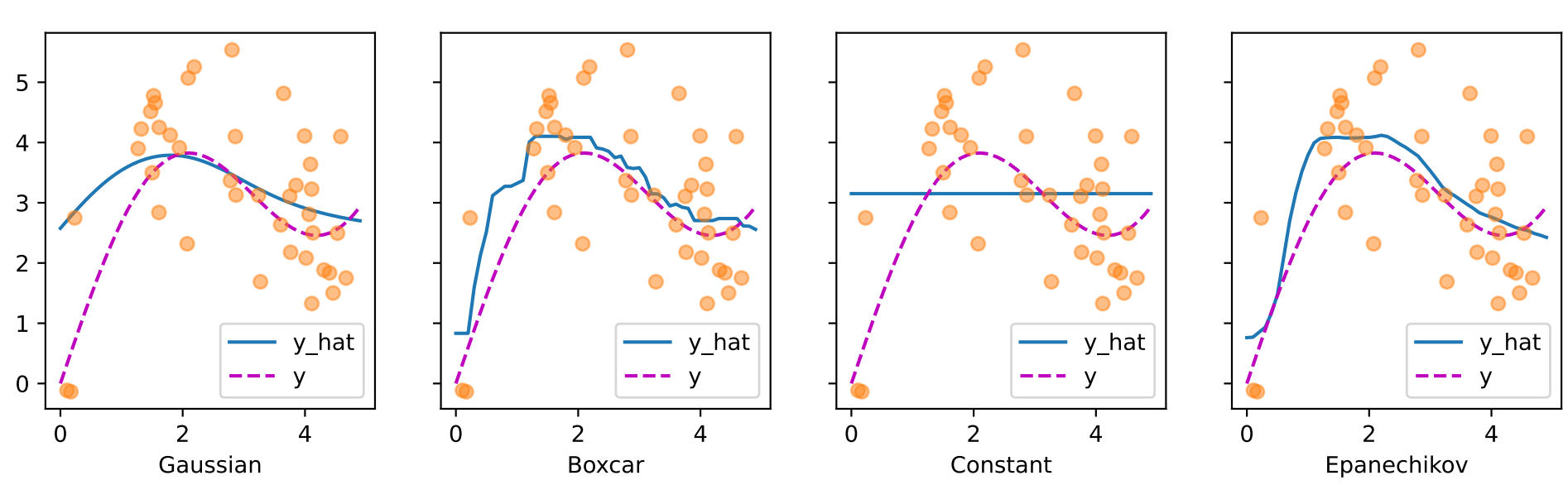

The first thing that stands out is that all three nontrivial kernels (Gaussian, Boxcar, and Epanechikov) produce fairly workable estimates that are not too far from the true function. Only the constant kernel that leads to the trivial estimate $f(x) = \frac{1}{n} \sum_i y_i$ produces a rather unrealistic result. Let's inspect the attention weighting a bit more closely:

```
plot(x_train, y_train, x_val, y_val, kernels, names, attention=True)
```

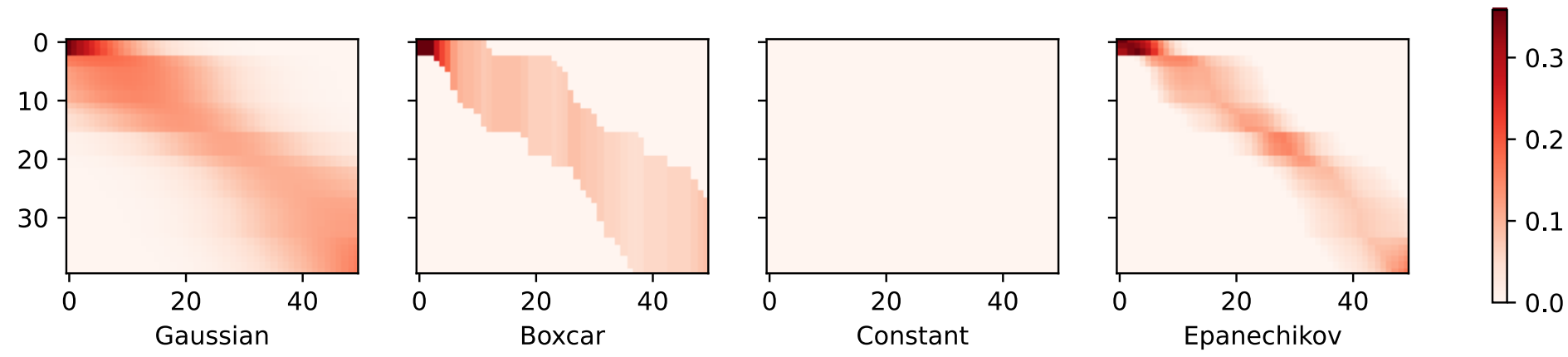

The visualization clearly shows why the estimates for Gaussian, Boxcar, and Epanechikov are very similar: after all, they are derived from very similar attention weights, despite the different functional form of the kernel. This raises the question as to whether this is always the case.

### 11.2.3 Adapting Attention Pooling

We could replace the Gaussian kernel with one of a different width. That is, we could use $\alpha(\mathbf{q}, \mathbf{k}) = \exp\left(-\frac{1}{2\sigma^2} \|\mathbf{q} - \mathbf{k}\|^2\right)$ where $\sigma^2$ determines the width of the kernel. Let's see whether this affects the outcomes.

```
sigmas = (0.1, 0.2, 0.5, 1)
names = ['Sigma ' + str(sigma) for sigma in sigmas]

def gaussian_with_width(sigma):
    return (lambda x: torch.exp(-x**2 / (2*sigma**2)))
```

```
kernels = [gaussian_with_width(sigma) for sigma in sigmas]
plot(x_train, y_train, x_val, y_val, kernels, names)
```

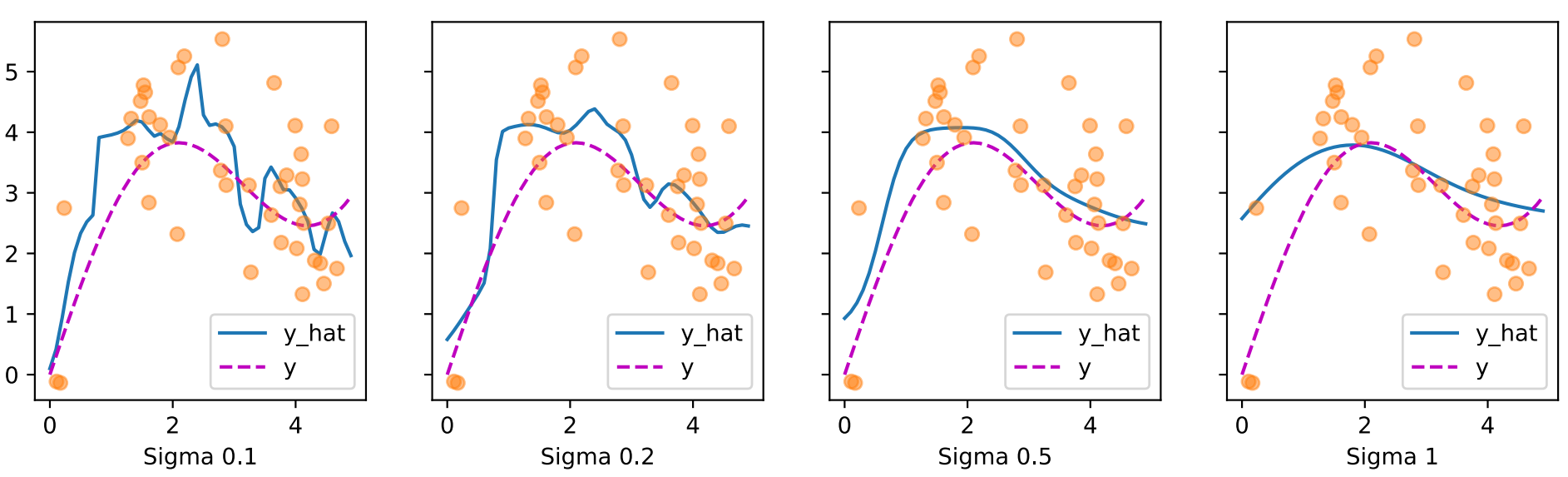

Clearly, the narrower the kernel, the less smooth the estimate. At the same time, it adapts better to the local variations. Let's look at the corresponding attention weights.

```
plot(x_train, y_train, x_val, y_val, kernels, names, attention=True)
```

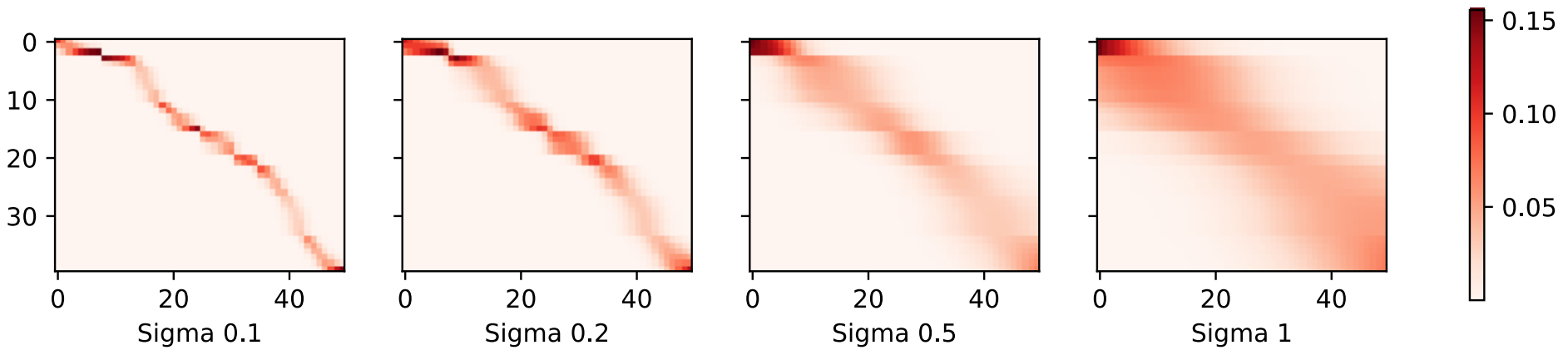

As we would expect, the narrower the kernel, the narrower the range of large attention weights. It is also clear that picking the same width might not be ideal. In fact, Silverman (1986) proposed a heuristic that depends on the local density. Many more such "tricks" have been proposed. For instance, Norelli *et al.* (2022) used a similar nearest-neighbor interpolation technique for designing cross-modal image and text representations.

The astute reader might wonder why we are providing this deep dive for a method that is over half a century old. First, it is one of the earliest precursors of modern attention mechanisms. Second, it is great for visualization. Third, and just as importantly, it demonstrates the limits of hand-crafted attention mechanisms. A much better strategy is to *learn* the mechanism, by learning the representations for queries and keys. This is what we will embark on in the following sections.

## 11.2.4 Summary

Nadaraya–Watson kernel regression is an early precursor of the current attention mechanisms. It can be used directly with little to no training or tuning, either for classification or regression.

The attention weight is assigned according to the similarity (or distance) between query and key, and according to how many similar observations are available.

### 11.2.5 Exercises

1. Parzen windows density estimates are given by $\hat{p}(\mathbf{x}) = \frac{1}{n} \sum_i k(\mathbf{x}, \mathbf{x}_i)$. Prove that for binary classification the function $\hat{p}(\mathbf{x}, y = 1) - \hat{p}(\mathbf{x}, y = -1)$, as obtained by Parzen windows is equivalent to Nadaraya–Watson classification.

2. Implement stochastic gradient descent to learn a good value for kernel widths in Nadaraya–Watson regression.

    1. What happens if you just use the above estimates to minimize $(f(\mathbf{x_i}) - y_i)^2$ directly? Hint: $y_i$ is part of the terms used to compute $f$.

    2. Remove $(\mathbf{x}_i, y_i)$ from the estimate for $f(\mathbf{x}_i)$ and optimize over the kernel widths. Do you still observe overfitting?

3. Assume that all $\mathbf{x}$ lie on the unit sphere, i.e., all satisfy $\|\mathbf{x}\| = 1$. Can you simplify the $\|\mathbf{x} - \mathbf{x}_i\|^2$ term in the exponential? Hint: we will later see that this is very closely related to dot product attention.

4. Recall that Mack and Silverman (1982) proved that Nadaraya–Watson estimation is consistent. How quickly should you reduce the scale for the attention mechanism as you get more data? Provide some intuition for your answer. Does it depend on the dimensionality of the data? How?

Discussions[156].

## 11.3 Attention Scoring Functions

In Section 11.2, we used a number of different distance-based kernels, including a Gaussian kernel to model interactions between queries and keys. As it turns out, distance functions are slightly more expensive to compute than dot products. As such, with the softmax operation to ensure nonnegative attention weights, much of the work has gone into *attention scoring functions* $a$ in (11.1.3) and Fig. 11.3.1 that are simpler to compute.

```
import math
import torch
from torch import nn
from d2l import torch as d2l
```

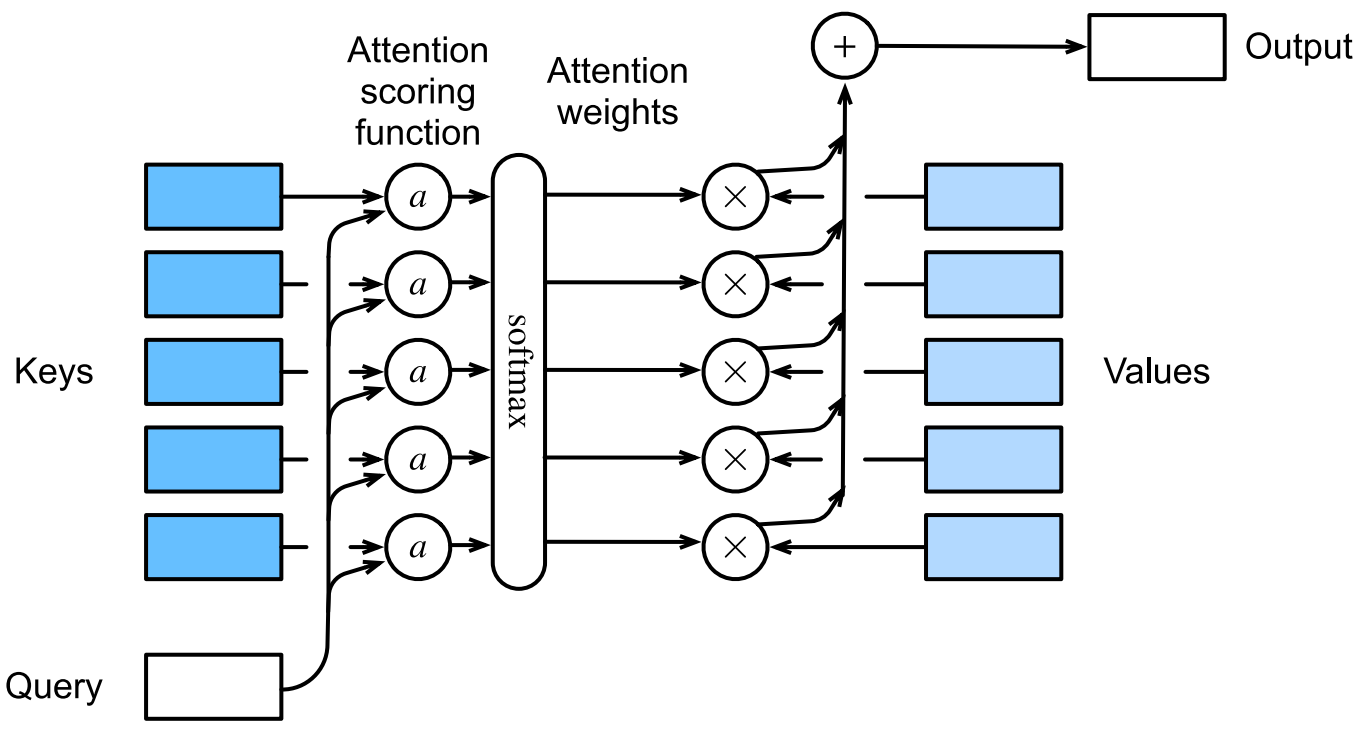

Fig. 11.3.1 Computing the output of attention pooling as a weighted average of values, where weights are computed with the attention scoring function $a$ and the softmax operation.

## 11.3.1 Dot Product Attention

Let's review the attention function (without exponentiation) from the Gaussian kernel for a moment:

$$a(\mathbf{q}, \mathbf{k}_i) = -\frac{1}{2}\|\mathbf{q} - \mathbf{k}_i\|^2 = \mathbf{q}^\top \mathbf{k}_i - \frac{1}{2}\|\mathbf{k}_i\|^2 - \frac{1}{2}\|\mathbf{q}\|^2. \tag{11.3.1}$$

First, note that the final term depends on $\mathbf{q}$ only. As such it is identical for all $(\mathbf{q}, \mathbf{k}_i)$ pairs. Normalizing the attention weights to 1, as is done in (11.1.3), ensures that this term disappears entirely. Second, note that both batch and layer normalization (to be discussed later) lead to activations that have well-bounded, and often constant, norms $\|\mathbf{k}_i\|$. This is the case, for instance, whenever the keys $\mathbf{k}_i$ were generated by a layer norm. As such, we can drop it from the definition of $a$ without any major change in the outcome.

Last, we need to keep the order of magnitude of the arguments in the exponential function under control. Assume that all the elements of the query $\mathbf{q} \in \mathbb{R}^d$ and the key $\mathbf{k}_i \in \mathbb{R}^d$ are independent and identically drawn random variables with zero mean and unit variance. The dot product between both vectors has zero mean and a variance of $d$. To ensure that the variance of the dot product still remains 1 regardless of vector length, we use the *scaled dot product attention* scoring function. That is, we rescale the dot product by $1/\sqrt{d}$. We thus arrive at the first commonly used attention function that is used, e.g., in Transformers (Vaswani *et al.*, 2017):

$$a(\mathbf{q}, \mathbf{k}_i) = \mathbf{q}^\top \mathbf{k}_i / \sqrt{d}. \tag{11.3.2}$$

Note that attention weights $\alpha$ still need normalizing. We can simplify this further via (11.1.3) by using the softmax operation:

$$\alpha(\mathbf{q}, \mathbf{k}_i) = \mathrm{softmax}(a(\mathbf{q}, \mathbf{k}_i)) = \frac{\exp(\mathbf{q}^\top \mathbf{k}_i / \sqrt{d})}{\sum_{j=1} \exp(\mathbf{q}^\top \mathbf{k}_j / \sqrt{d})}. \tag{11.3.3}$$

As it turns out, all popular attention mechanisms use the softmax, hence we will limit ourselves to that in the remainder of this chapter.

## 11.3.2 Convenience Functions

We need a few functions to make the attention mechanism efficient to deploy. This includes tools for dealing with strings of variable lengths (common for natural language processing) and tools for efficient evaluation on minibatches (batch matrix multiplication).

### Masked Softmax Operation

One of the most popular applications of the attention mechanism is to sequence models. Hence we need to be able to deal with sequences of different lengths. In some cases, such sequences may end up in the same minibatch, necessitating padding with dummy tokens for shorter sequences (see Section 10.5 for an example). These special tokens do not carry meaning. For instance, assume that we have the following three sentences:

```
Dive  into  Deep    Learning
Learn to    code    <blank>
Hello world <blank> <blank>
```

Since we do not want blanks in our attention model we simply need to limit $\sum_{i=1}^{n} \alpha(\mathbf{q}, \mathbf{k}_i)\mathbf{v}_i$ to $\sum_{i=1}^{l} \alpha(\mathbf{q}, \mathbf{k}_i)\mathbf{v}_i$ for however long, $l \leq n$, the actual sentence is. Since it is such a common problem, it has a name: the *masked softmax operation*.

Let's implement it. Actually, the implementation cheats ever so slightly by setting the values of $\mathbf{v}_i$, for $i > l$, to zero. Moreover, it sets the attention weights to a large negative number, such as $-10^6$, in order to make their contribution to gradients and values vanish in practice. This is done since linear algebra kernels and operators are heavily optimized for GPUs and it is faster to be slightly wasteful in computation rather than to have code with conditional (if then else) statements.

```python
def masked_softmax(X, valid_lens):  #@save
    """Perform softmax operation by masking elements on the last axis."""
    # X: 3D tensor, valid_lens: 1D or 2D tensor
    def _sequence_mask(X, valid_len, value=0):
        maxlen = X.size(1)
        mask = torch.arange((maxlen), dtype=torch.float32,
                            device=X.device)[None, :] < valid_len[:, None]
        X[~mask] = value
        return X

    if valid_lens is None:
        return nn.functional.softmax(X, dim=-1)
    else:
        shape = X.shape
        if valid_lens.dim() == 1:
            valid_lens = torch.repeat_interleave(valid_lens, shape[1])
        else:
            valid_lens = valid_lens.reshape(-1)
        # On the last axis, replace masked elements with a very large negative
```

```
    # value, whose exponentiation outputs 0
    X = _sequence_mask(X.reshape(-1, shape[-1]), valid_lens, value=-1e6)
    return nn.functional.softmax(X.reshape(shape), dim=-1)
```

To illustrate how this function works, consider a minibatch of two examples of size $2 \times 4$, where their valid lengths are 2 and 3, respectively. As a result of the masked softmax operation, values beyond the valid lengths for each pair of vectors are all masked as zero.

```
masked_softmax(torch.rand(2, 2, 4), torch.tensor([2, 3]))
```

```
tensor([[[0.5496, 0.4504, 0.0000, 0.0000],
         [0.5542, 0.4458, 0.0000, 0.0000]],

        [[0.3448, 0.2301, 0.4251, 0.0000],
         [0.3145, 0.3813, 0.3041, 0.0000]]])
```

If we need more fine-grained control to specify the valid length for each of the two vectors of every example, we simply use a two-dimensional tensor of valid lengths. This yields:

```
masked_softmax(torch.rand(2, 2, 4), torch.tensor([[1, 3], [2, 4]]))
```

```
tensor([[[1.0000, 0.0000, 0.0000, 0.0000],
         [0.2975, 0.4037, 0.2989, 0.0000]],

        [[0.5417, 0.4583, 0.0000, 0.0000],
         [0.1761, 0.2262, 0.3378, 0.2598]]])
```

### Batch Matrix Multiplication

Another commonly used operation is to multiply batches of matrices by one another. This comes in handy when we have minibatches of queries, keys, and values. More specifically, assume that

$$\begin{aligned}
\mathbf{Q} &= [\mathbf{Q}_1, \mathbf{Q}_2, \ldots, \mathbf{Q}_n] \in \mathbb{R}^{n \times a \times b}, \\
\mathbf{K} &= [\mathbf{K}_1, \mathbf{K}_2, \ldots, \mathbf{K}_n] \in \mathbb{R}^{n \times b \times c}.
\end{aligned} \tag{11.3.4}$$

Then the batch matrix multiplication (BMM) computes the elementwise product

$$\mathrm{BMM}(\mathbf{Q}, \mathbf{K}) = [\mathbf{Q}_1 \mathbf{K}_1, \mathbf{Q}_2 \mathbf{K}_2, \ldots, \mathbf{Q}_n \mathbf{K}_n] \in \mathbb{R}^{n \times a \times c}. \tag{11.3.5}$$

Let's see this in action in a deep learning framework.

```
Q = torch.ones((2, 3, 4))
K = torch.ones((2, 4, 6))
d2l.check_shape(torch.bmm(Q, K), (2, 3, 6))
```

### 11.3.3 Scaled Dot Product Attention

Let's return to the dot product attention introduced in (11.3.2). In general, it requires that both the query and the key have the same vector length, say $d$, even though this can be addressed easily by replacing $\mathbf{q}^\top \mathbf{k}$ with $\mathbf{q}^\top \mathbf{M} \mathbf{k}$ where $\mathbf{M}$ is a matrix suitably chosen for translating between both spaces. For now assume that the dimensions match.

In practice, we often think of minibatches for efficiency, such as computing attention for $n$ queries and $m$ key-value pairs, where queries and keys are of length $d$ and values are of length $v$. The scaled dot product attention of queries $\mathbf{Q} \in \mathbb{R}^{n\times d}$, keys $\mathbf{K} \in \mathbb{R}^{m\times d}$, and values $\mathbf{V} \in \mathbb{R}^{m\times v}$ thus can be written as

$$\mathrm{softmax}\left(\frac{\mathbf{Q}\mathbf{K}^\top}{\sqrt{d}}\right)\mathbf{V} \in \mathbb{R}^{n\times v}. \tag{11.3.6}$$

Note that when applying this to a minibatch, we need the batch matrix multiplication introduced in (11.3.5). In the following implementation of the scaled dot product attention, we use dropout for model regularization.

```
class DotProductAttention(nn.Module):  #@save
    """Scaled dot product attention."""
    def __init__(self, dropout):
        super().__init__()
        self.dropout = nn.Dropout(dropout)

    # Shape of queries: (batch_size, no. of queries, d)
    # Shape of keys: (batch_size, no. of key-value pairs, d)
    # Shape of values: (batch_size, no. of key-value pairs, value dimension)
    # Shape of valid_lens: (batch_size,) or (batch_size, no. of queries)
    def forward(self, queries, keys, values, valid_lens=None):
        d = queries.shape[-1]
        # Swap the last two dimensions of keys with keys.transpose(1, 2)
        scores = torch.bmm(queries, keys.transpose(1, 2)) / math.sqrt(d)
        self.attention_weights = masked_softmax(scores, valid_lens)
        return torch.bmm(self.dropout(self.attention_weights), values)
```

To illustrate how the `DotProductAttention` class works, we use the same keys, values, and valid lengths from the earlier toy example for additive attention. For the purpose of our example we assume that we have a minibatch size of 2, a total of 10 keys and values, and that the dimensionality of the values is 4. Lastly, we assume that the valid length per observation is 2 and 6 respectively. Given that, we expect the output to be a $2 \times 1 \times 4$ tensor, i.e., one row per example of the minibatch.

```
queries = torch.normal(0, 1, (2, 1, 2))
keys = torch.normal(0, 1, (2, 10, 2))
values = torch.normal(0, 1, (2, 10, 4))
valid_lens = torch.tensor([2, 6])

attention = DotProductAttention(dropout=0.5)
attention.eval()
d2l.check_shape(attention(queries, keys, values, valid_lens), (2, 1, 4))
```

Let's check whether the attention weights actually vanish for anything beyond the second and sixth column respectively (because of setting the valid length to 2 and 6).

```
d2l.show_heatmaps(attention.attention_weights.reshape((1, 1, 2, 10)),
                  xlabel='Keys', ylabel='Queries')
```

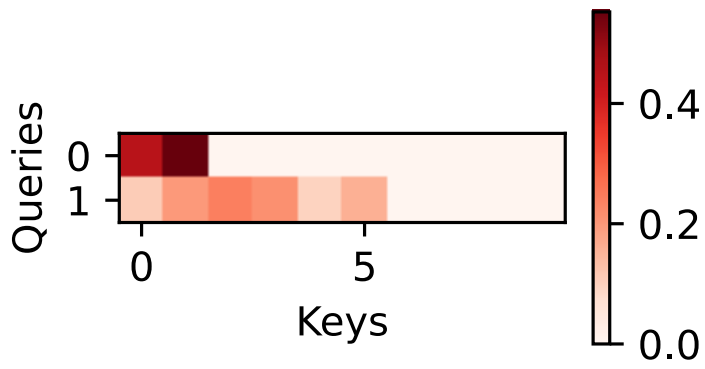

## 11.3.4 Additive Attention

When queries $\mathbf{q}$ and keys $\mathbf{k}$ are vectors of different dimension, we can either use a matrix to address the mismatch via $\mathbf{q}^\top \mathbf{M} \mathbf{k}$, or we can use additive attention as the scoring function. Another benefit is that, as its name indicates, the attention is additive. This can lead to some minor computational savings. Given a query $\mathbf{q} \in \mathbb{R}^q$ and a key $\mathbf{k} \in \mathbb{R}^k$, the *additive attention* scoring function (Bahdanau *et al.*, 2014) is given by

$$a(\mathbf{q}, \mathbf{k}) = \mathbf{w}_v^\top \tanh(\mathbf{W}_q \mathbf{q} + \mathbf{W}_k \mathbf{k}) \in \mathbb{R}, \tag{11.3.7}$$

where $\mathbf{W}_q \in \mathbb{R}^{h \times q}$, $\mathbf{W}_k \in \mathbb{R}^{h \times k}$, and $\mathbf{w}_v \in \mathbb{R}^h$ are the learnable parameters. This term is then fed into a softmax to ensure both nonnegativity and normalization. An equivalent interpretation of (11.3.7) is that the query and key are concatenated and fed into an MLP with a single hidden layer. Using $\tanh$ as the activation function and disabling bias terms, we implement additive attention as follows:

```
class AdditiveAttention(nn.Module):  #@save
    """Additive attention."""
    def __init__(self, num_hiddens, dropout, **kwargs):
        super(AdditiveAttention, self).__init__(**kwargs)
        self.W_k = nn.LazyLinear(num_hiddens, bias=False)
        self.W_q = nn.LazyLinear(num_hiddens, bias=False)
        self.w_v = nn.LazyLinear(1, bias=False)
        self.dropout = nn.Dropout(dropout)

    def forward(self, queries, keys, values, valid_lens):
        queries, keys = self.W_q(queries), self.W_k(keys)
        # After dimension expansion, shape of queries: (batch_size, no. of
        # queries, 1, num_hiddens) and shape of keys: (batch_size, 1, no. of
        # key-value pairs, num_hiddens). Sum them up with broadcasting
        features = queries.unsqueeze(2) + keys.unsqueeze(1)
        features = torch.tanh(features)
        # There is only one output of self.w_v, so we remove the last
        # one-dimensional entry from the shape. Shape of scores: (batch_size,
        # no. of queries, no. of key-value pairs)
```

(continued from previous page)

```
        scores = self.w_v(features).squeeze(-1)
        self.attention_weights = masked_softmax(scores, valid_lens)
        # Shape of values: (batch_size, no. of key-value pairs, value
        # dimension)
        return torch.bmm(self.dropout(self.attention_weights), values)
```

Let's see how `AdditiveAttention` works. In our toy example we pick queries, keys and values of size $(2, 1, 20)$, $(2, 10, 2)$ and $(2, 10, 4)$, respectively. This is identical to our choice for `DotProductAttention`, except that now the queries are 20-dimensional. Likewise, we pick $(2, 6)$ as the valid lengths for the sequences in the minibatch.

```
queries = torch.normal(0, 1, (2, 1, 20))

attention = AdditiveAttention(num_hiddens=8, dropout=0.1)
attention.eval()
d2l.check_shape(attention(queries, keys, values, valid_lens), (2, 1, 4))
```

When reviewing the attention function we see a behavior that is qualitatively quite similar to that of `DotProductAttention`. That is, only terms within the chosen valid length $(2, 6)$ are nonzero.

```
d2l.show_heatmaps(attention.attention_weights.reshape((1, 1, 2, 10)),
                  xlabel='Keys', ylabel='Queries')
```

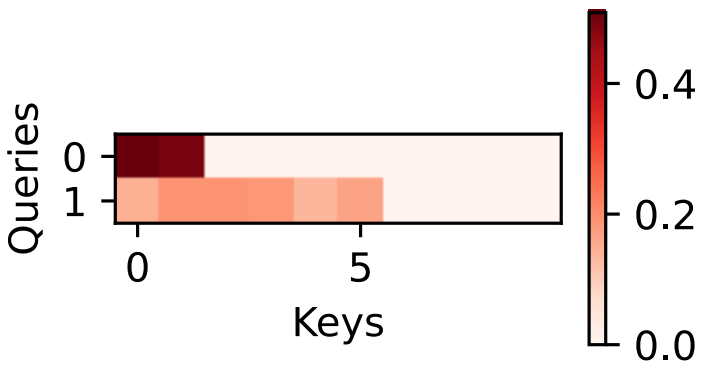

## 11.3.5 Summary

In this section we introduced the two key attention scoring functions: dot product and additive attention. They are effective tools for aggregating across sequences of variable length. In particular, the dot product attention is the mainstay of modern Transformer architectures. When queries and keys are vectors of different lengths, we can use the additive attention scoring function instead. Optimizing these layers is one of the key areas of advance in recent years. For instance, NVIDIA's Transformer Library[157] and Megatron (Shoeybi *et al.*, 2019) crucially rely on efficient variants of the attention mechanism. We will dive into this in quite a bit more detail as we review Transformers in later sections.

## 11.3.6 Exercises

1. Implement distance-based attention by modifying the `DotProductAttention` code. Note that you only need the squared norms of the keys $\|\mathbf{k}_i\|^2$ for an efficient implementation.
2. Modify the dot product attention to allow for queries and keys of different dimensionalities by employing a matrix to adjust dimensions.
3. How does the computational cost scale with the dimensionality of the keys, queries, values, and their number? What about the memory bandwidth requirements?

Discussions[158].

## 11.4 The Bahdanau Attention Mechanism

When we encountered machine translation in Section 10.7, we designed an encoder–decoder architecture for sequence-to-sequence learning based on two RNNs (Sutskever *et al.*, 2014). Specifically, the RNN encoder transforms a variable-length sequence into a *fixed-shape* context variable. Then, the RNN decoder generates the output (target) sequence token by token based on the generated tokens and the context variable.

Recall Fig. 10.7.2 which we repeat (Fig. 11.4.1) with some additional detail. Conventionally, in an RNN all relevant information about a source sequence is translated into some internal *fixed-dimensional* state representation by the encoder. It is this very state that is used by the decoder as the complete and exclusive source of information for generating the translated sequence. In other words, the sequence-to-sequence mechanism treats the intermediate state as a sufficient statistic of whatever string might have served as input.

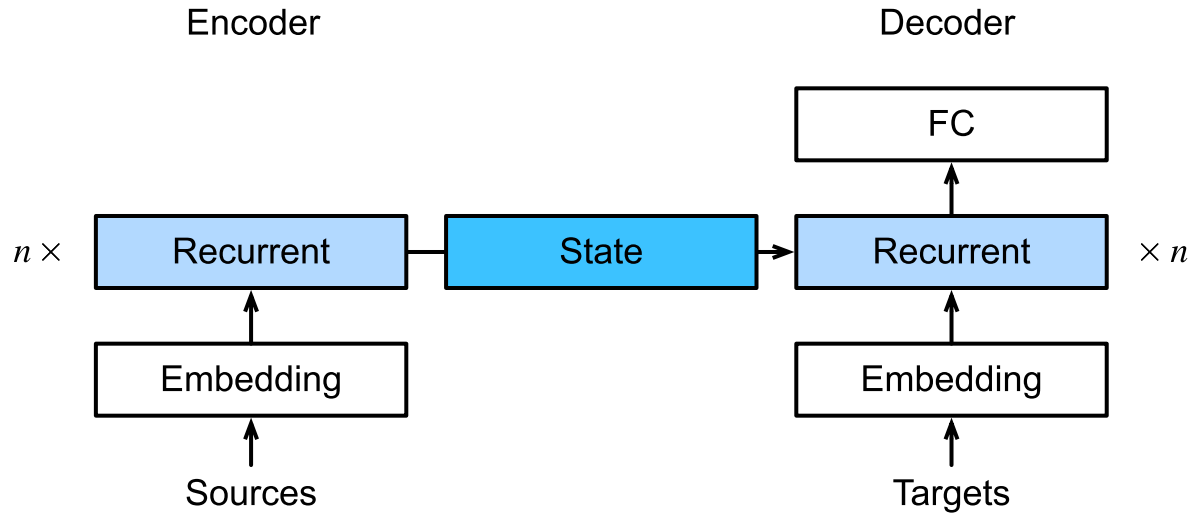

Fig. 11.4.1 Sequence-to-sequence model. The state, as generated by the encoder, is the only piece of information shared between the encoder and the decoder.

While this is quite reasonable for short sequences, it is clear that it is infeasible for long ones, such as a book chapter or even just a very long sentence. After all, before too long there will simply not be enough "space" in the intermediate representation to store all that is important in the source sequence. Consequently the decoder will fail to translate long and complex sentences. One of the first to encounter this was Graves (2013) who tried to design an RNN to generate handwritten text. Since the source text has arbitrary length they designed a differentiable attention model to align text characters with the much longer pen trace, where the alignment moves only in one direction. This, in turn, draws on decoding algorithms in speech recognition, e.g., hidden Markov models (Rabiner and Juang, 1993).

Inspired by the idea of learning to align, Bahdanau *et al.* (2014) proposed a differentiable attention model *without* the unidirectional alignment limitation. When predicting a token, if not all the input tokens are relevant, the model aligns (or attends) only to parts of the input sequence that are deemed relevant to the current prediction. This is then used to update the current state before generating the next token. While quite innocuous in its description, this *Bahdanau attention mechanism* has arguably turned into one of the most influential ideas of the past decade in deep learning, giving rise to Transformers (Vaswani *et al.*, 2017) and many related new architectures.

```
import torch
from torch import nn
from d2l import torch as d2l
```

## 11.4.1 Model

We follow the notation introduced by the sequence-to-sequence architecture of Section 10.7, in particular (10.7.3). The key idea is that instead of keeping the state, i.e., the context variable $\mathbf{c}$ summarizing the source sentence, as fixed, we dynamically update it, as a function of both the original text (encoder hidden states $\mathbf{h}_t$) and the text that was already generated (decoder hidden states $\mathbf{s}_{t'-1}$). This yields $\mathbf{c}_{t'}$, which is updated after any decoding time step $t'$. Suppose that the input sequence is of length $T$. In this case the context variable is the output of attention pooling:

$$\mathbf{c}_{t'} = \sum_{t=1}^{T} \alpha(\mathbf{s}_{t'-1}, \mathbf{h}_t)\mathbf{h}_t. \tag{11.4.1}$$

We used $\mathbf{s}_{t'-1}$ as the query, and $\mathbf{h}_t$ as both the key and the value. Note that $\mathbf{c}_{t'}$ is then used to generate the state $\mathbf{s}_{t'}$ and to generate a new token: see (10.7.3). In particular, the attention weight $\alpha$ is computed as in (11.3.3) using the additive attention scoring function defined by (11.3.7). This RNN encoder–decoder architecture using attention is depicted in Fig. 11.4.2. Note that later this model was modified so as to include the already generated tokens in the decoder as further context (i.e., the attention sum does not stop at $T$ but rather it proceeds up to $t' - 1$). For instance, see Chan *et al.* (2015) for a description of this strategy, as applied to speech recognition.

## 11.4.2 Defining the Decoder with Attention

To implement the RNN encoder–decoder with attention, we only need to redefine the decoder (omitting the generated symbols from the attention function simplifies the design). Let's begin with the base interface for decoders with attention by defining the quite unsurprisingly named `AttentionDecoder` class.

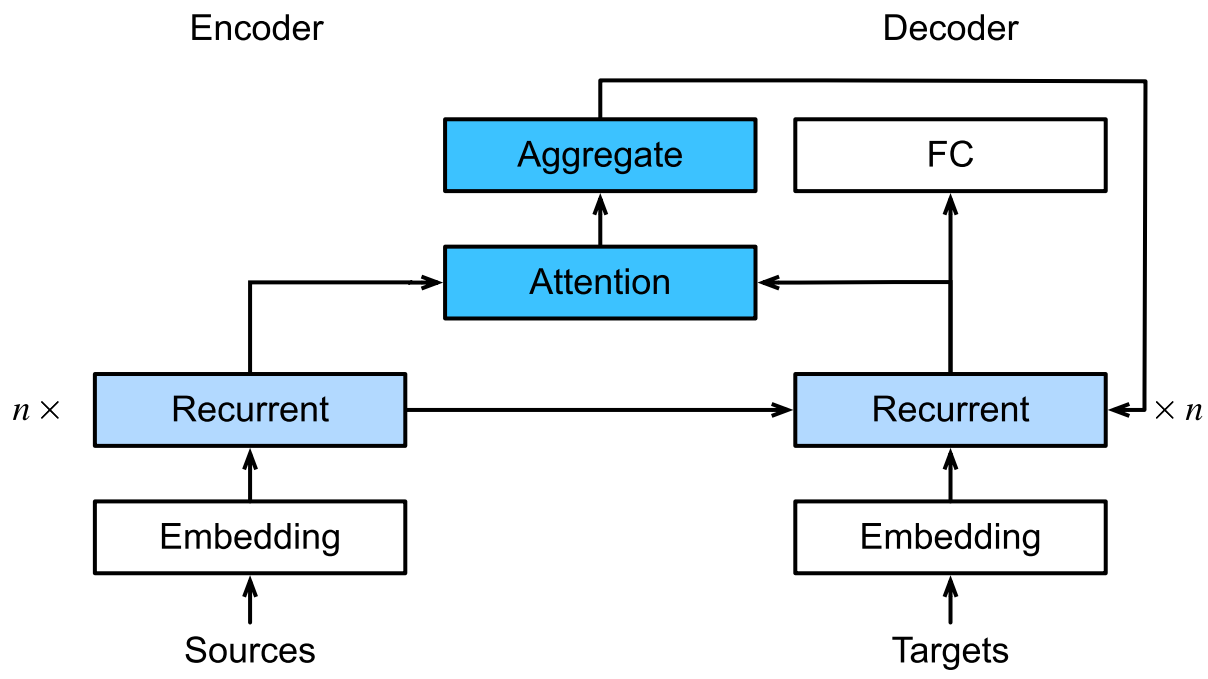

Fig. 11.4.2 Layers in an RNN encoder–decoder model with the Bahdanau attention mechanism.

```
class AttentionDecoder(d2l.Decoder):  #@save
    """The base attention-based decoder interface."""
    def __init__(self):
        super().__init__()

    @property
    def attention_weights(self):
        raise NotImplementedError
```

We need to implement the RNN decoder in the `Seq2SeqAttentionDecoder` class. The state of the decoder is initialized with (i) the hidden states of the last layer of the encoder at all time steps, used as keys and values for attention; (ii) the hidden state of the encoder at all layers at the final time step, which serves to initialize the hidden state of the decoder; and (iii) the valid length of the encoder, to exclude the padding tokens in attention pooling. At each decoding time step, the hidden state of the final layer of the decoder, obtained at the previous time step, is used as the query of the attention mechanism. Both the output of the attention mechanism and the input embedding are concatenated to serve as the input of the RNN decoder.

```
class Seq2SeqAttentionDecoder(AttentionDecoder):
    def __init__(self, vocab_size, embed_size, num_hiddens, num_layers,
                 dropout=0):
        super().__init__()
        self.attention = d2l.AdditiveAttention(num_hiddens, dropout)
        self.embedding = nn.Embedding(vocab_size, embed_size)
        self.rnn = nn.GRU(
            embed_size + num_hiddens, num_hiddens, num_layers,
            dropout=dropout)
        self.dense = nn.LazyLinear(vocab_size)
        self.apply(d2l.init_seq2seq)

    def init_state(self, enc_outputs, enc_valid_lens):
        # Shape of outputs: (num_steps, batch_size, num_hiddens).
```

```
        # Shape of hidden_state: (num_layers, batch_size, num_hiddens)
        outputs, hidden_state = enc_outputs
        return (outputs.permute(1, 0, 2), hidden_state, enc_valid_lens)

    def forward(self, X, state):
        # Shape of enc_outputs: (batch_size, num_steps, num_hiddens).
        # Shape of hidden_state: (num_layers, batch_size, num_hiddens)
        enc_outputs, hidden_state, enc_valid_lens = state
        # Shape of the output X: (num_steps, batch_size, embed_size)
        X = self.embedding(X).permute(1, 0, 2)
        outputs, self._attention_weights = [], []
        for x in X:
            # Shape of query: (batch_size, 1, num_hiddens)
            query = torch.unsqueeze(hidden_state[-1], dim=1)
            # Shape of context: (batch_size, 1, num_hiddens)
            context = self.attention(
                query, enc_outputs, enc_outputs, enc_valid_lens)
            # Concatenate on the feature dimension
            x = torch.cat((context, torch.unsqueeze(x, dim=1)), dim=-1)
            # Reshape x as (1, batch_size, embed_size + num_hiddens)
            out, hidden_state = self.rnn(x.permute(1, 0, 2), hidden_state)
            outputs.append(out)
            self._attention_weights.append(self.attention.attention_weights)
        # After fully connected layer transformation, shape of outputs:
        # (num_steps, batch_size, vocab_size)
        outputs = self.dense(torch.cat(outputs, dim=0))
        return outputs.permute(1, 0, 2), [enc_outputs, hidden_state,
                                          enc_valid_lens]

    @property
    def attention_weights(self):
        return self._attention_weights
```

In the following, we test the implemented decoder with attention using a minibatch of four sequences, each of which are seven time steps long.

```
vocab_size, embed_size, num_hiddens, num_layers = 10, 8, 16, 2
batch_size, num_steps = 4, 7
encoder = d2l.Seq2SeqEncoder(vocab_size, embed_size, num_hiddens, num_layers)
decoder = Seq2SeqAttentionDecoder(vocab_size, embed_size, num_hiddens,
                                  num_layers)
X = torch.zeros((batch_size, num_steps), dtype=torch.long)
state = decoder.init_state(encoder(X), None)
output, state = decoder(X, state)
d2l.check_shape(output, (batch_size, num_steps, vocab_size))
d2l.check_shape(state[0], (batch_size, num_steps, num_hiddens))
d2l.check_shape(state[1][0], (batch_size, num_hiddens))
```

### 11.4.3 Training

Now that we specified the new decoder we can proceed analogously to Section 10.7.6: specify the hyperparameters, instantiate a regular encoder and a decoder with attention, and train this model for machine translation.

```
data = d2l.MTFraEng(batch_size=128)
embed_size, num_hiddens, num_layers, dropout = 256, 256, 2, 0.2
encoder = d2l.Seq2SeqEncoder(
    len(data.src_vocab), embed_size, num_hiddens, num_layers, dropout)
decoder = Seq2SeqAttentionDecoder(
    len(data.tgt_vocab), embed_size, num_hiddens, num_layers, dropout)
model = d2l.Seq2Seq(encoder, decoder, tgt_pad=data.tgt_vocab['<pad>'],
                    lr=0.005)
trainer = d2l.Trainer(max_epochs=30, gradient_clip_val=1, num_gpus=1)
trainer.fit(model, data)
```

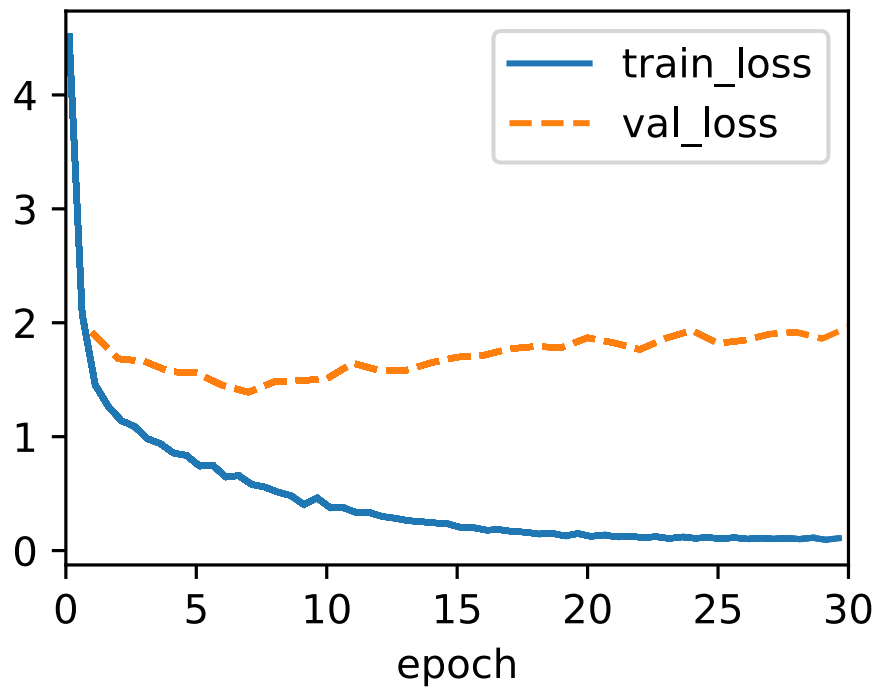

After the model is trained, we use it to translate a few English sentences into French and compute their BLEU scores.

```
engs = ['go .', 'i lost .', 'he\'s calm .', 'i\'m home .']
fras = ['va !', 'j\'ai perdu .', 'il est calme .', 'je suis chez moi .']
preds, _ = model.predict_step(
    data.build(engs, fras), d2l.try_gpu(), data.num_steps)
for en, fr, p in zip(engs, fras, preds):
    translation = []
    for token in data.tgt_vocab.to_tokens(p):
        if token == '<eos>':
            break
        translation.append(token)
    print(f'{en} => {translation}, bleu,'
          f'{d2l.bleu(" ".join(translation), fr, k=2):.3f}')
```

```
go . => ['va', '!'], bleu,1.000
i lost . => ["j'ai", 'perdu', '.'], bleu,1.000
he's calm . => ['je', 'suis', 'mouillé', '.'], bleu,0.000
i'm home . => ['je', 'suis', 'chez', 'moi', '.'], bleu,1.000
```

Let's visualize the attention weights when translating the last English sentence. We see that each query assigns non-uniform weights over key–value pairs. It shows that at each decoding step, different parts of the input sequences are selectively aggregated in the attention pooling.

```
_, dec_attention_weights = model.predict_step(
    data.build([engs[-1]], [fras[-1]]), d2l.try_gpu(), data.num_steps, True)
attention_weights = torch.cat(
    [step[0][0][0] for step in dec_attention_weights], 0)
attention_weights = attention_weights.reshape((1, 1, -1, data.num_steps))
```

```
# Plus one to include the end-of-sequence token
d2l.show_heatmaps(
    attention_weights[:, :, :, :len(engs[-1].split()) + 1].cpu(),
    xlabel='Key positions', ylabel='Query positions')
```

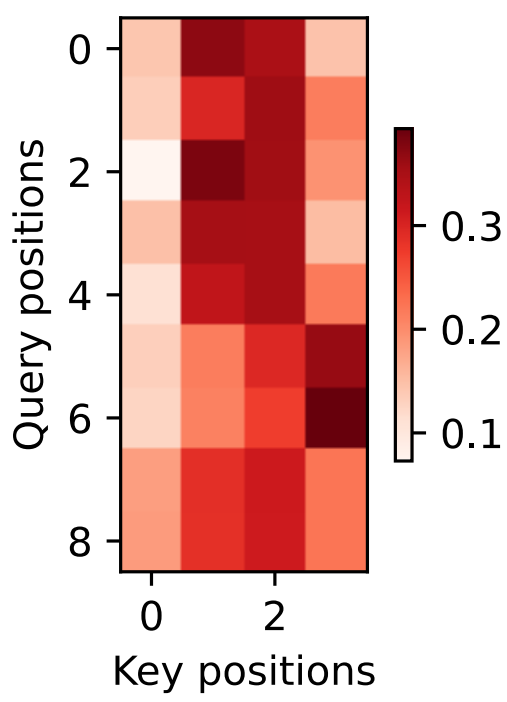

## 11.4.4 Summary

When predicting a token, if not all the input tokens are relevant, the RNN encoder–decoder with the Bahdanau attention mechanism selectively aggregates different parts of the input sequence. This is achieved by treating the state (context variable) as an output of additive attention pooling. In the RNN encoder–decoder, the Bahdanau attention mechanism treats the decoder hidden state at the previous time step as the query, and the encoder hidden states at all the time steps as both the keys and values.

## 11.4.5 Exercises

1. Replace GRU with LSTM in the experiment.
2. Modify the experiment to replace the additive attention scoring function with the scaled dot-product. How does it influence the training efficiency?

Discussions[159].

# 11.5 Multi-Head Attention

In practice, given the same set of queries, keys, and values we may want our model to combine knowledge from different behaviors of the same attention mechanism, such as capturing dependencies of various ranges (e.g., shorter-range vs. longer-range) within a sequence. Thus, it may be beneficial to allow our attention mechanism to jointly use different representation subspaces of queries, keys, and values.

To this end, instead of performing a single attention pooling, queries, keys, and values can be transformed with $h$ independently learned linear projections. Then these $h$ projected queries, keys, and values are fed into attention pooling in parallel. In the end, $h$ attention-pooling outputs are concatenated and transformed with another learned linear projection to produce the final output. This design is called *multi-head attention*, where each of the $h$ attention pooling outputs is a *head* (Vaswani *et al.*, 2017). Using fully connected layers to perform learnable linear transformations, Fig. 11.5.1 describes multi-head attention.

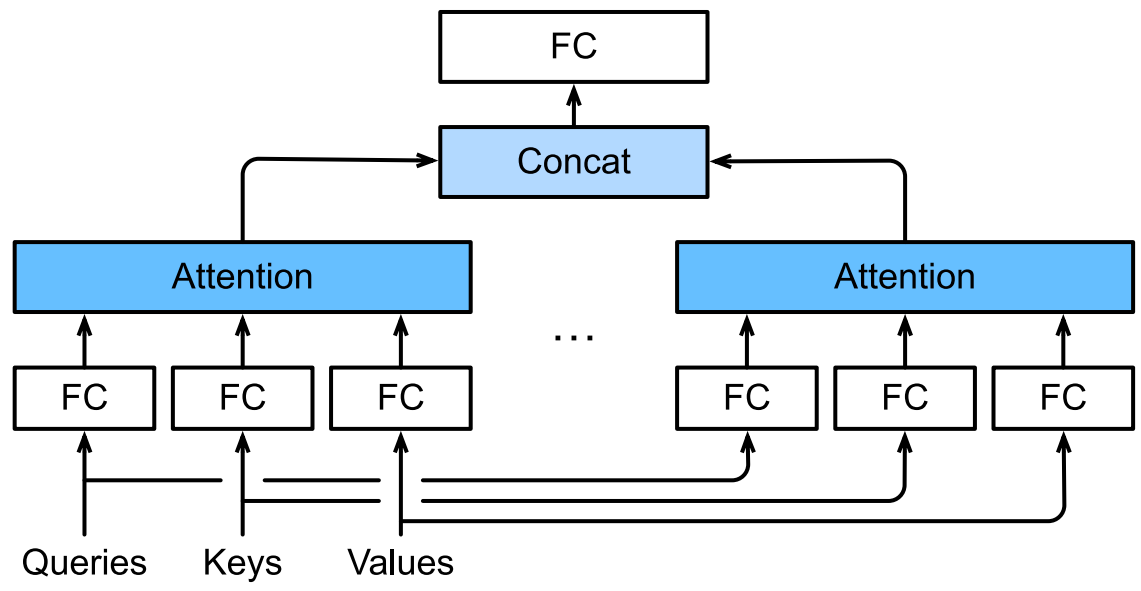

Fig. 11.5.1 Multi-head attention, where multiple heads are concatenated then linearly transformed.

```
import math
import torch
from torch import nn
from d2l import torch as d2l
```

## 11.5.1 Model

Before providing the implementation of multi-head attention, let's formalize this model mathematically. Given a query $\mathbf{q} \in \mathbb{R}^{d_q}$, a key $\mathbf{k} \in \mathbb{R}^{d_k}$, and a value $\mathbf{v} \in \mathbb{R}^{d_v}$, each attention head $\mathbf{h}_i$ $(i = 1, \ldots, h)$ is computed as

$$\mathbf{h}_i = f(\mathbf{W}_i^{(q)}\mathbf{q}, \mathbf{W}_i^{(k)}\mathbf{k}, \mathbf{W}_i^{(v)}\mathbf{v}) \in \mathbb{R}^{p_v}, \tag{11.5.1}$$

where $\mathbf{W}_i^{(q)} \in \mathbb{R}^{p_q \times d_q}$, $\mathbf{W}_i^{(k)} \in \mathbb{R}^{p_k \times d_k}$, and $\mathbf{W}_i^{(v)} \in \mathbb{R}^{p_v \times d_v}$ are learnable parameters and $f$ is attention pooling, such as additive attention and scaled dot product attention in Section 11.3. The multi-head attention output is another linear transformation via learnable parameters $\mathbf{W}_o \in \mathbb{R}^{p_o \times h p_v}$ of the concatenation of $h$ heads:

$$\mathbf{W}_o \begin{bmatrix}\mathbf{h}_1\\\vdots\\\mathbf{h}_h\end{bmatrix} \in \mathbb{R}^{p_o}. \tag{11.5.2}$$

Based on this design, each head may attend to different parts of the input. More sophisticated functions than the simple weighted average can be expressed.

## 11.5.2 Implementation

In our implementation, we choose the scaled dot product attention for each head of the multi-head attention. To avoid significant growth of computational cost and parametrization cost, we set $p_q = p_k = p_v = p_o/h$. Note that $h$ heads can be computed in parallel if we set the number of outputs of linear transformations for the query, key, and value to $p_q h = p_k h = p_v h = p_o$. In the following implementation, $p_o$ is specified via the argument `num_hiddens`.

```
class MultiHeadAttention(d2l.Module):  #@save
    """Multi-head attention."""
    def __init__(self, num_hiddens, num_heads, dropout, bias=False, **kwargs):
        super().__init__()
        self.num_heads = num_heads
        self.attention = d2l.DotProductAttention(dropout)
        self.W_q = nn.LazyLinear(num_hiddens, bias=bias)
        self.W_k = nn.LazyLinear(num_hiddens, bias=bias)
        self.W_v = nn.LazyLinear(num_hiddens, bias=bias)
        self.W_o = nn.LazyLinear(num_hiddens, bias=bias)

    def forward(self, queries, keys, values, valid_lens):
        # Shape of queries, keys, or values:
        # (batch_size, no. of queries or key-value pairs, num_hiddens)
        # Shape of valid_lens: (batch_size,) or (batch_size, no. of queries)
        # After transposing, shape of output queries, keys, or values:
        # (batch_size * num_heads, no. of queries or key-value pairs,
        # num_hiddens / num_heads)
        queries = self.transpose_qkv(self.W_q(queries))
        keys = self.transpose_qkv(self.W_k(keys))
        values = self.transpose_qkv(self.W_v(values))

        if valid_lens is not None:
            # On axis 0, copy the first item (scalar or vector) for num_heads
            # times, then copy the next item, and so on
            valid_lens = torch.repeat_interleave(
                valid_lens, repeats=self.num_heads, dim=0)

        # Shape of output: (batch_size * num_heads, no. of queries,
        # num_hiddens / num_heads)
        output = self.attention(queries, keys, values, valid_lens)
        # Shape of output_concat: (batch_size, no. of queries, num_hiddens)
```

```
        output_concat = self.transpose_output(output)
        return self.W_o(output_concat)
```

To allow for parallel computation of multiple heads, the above `MultiHeadAttention` class uses two transposition methods as defined below. Specifically, the `transpose_output` method reverses the operation of the `transpose_qkv` method.

```
@d2l.add_to_class(MultiHeadAttention)  #@save
def transpose_qkv(self, X):
    """Transposition for parallel computation of multiple attention heads."""
    # Shape of input X: (batch_size, no. of queries or key-value pairs,
    # num_hiddens). Shape of output X: (batch_size, no. of queries or
    # key-value pairs, num_heads, num_hiddens / num_heads)
    X = X.reshape(X.shape[0], X.shape[1], self.num_heads, -1)
    # Shape of output X: (batch_size, num_heads, no. of queries or key-value
    # pairs, num_hiddens / num_heads)
    X = X.permute(0, 2, 1, 3)
    # Shape of output: (batch_size * num_heads, no. of queries or key-value
    # pairs, num_hiddens / num_heads)
    return X.reshape(-1, X.shape[2], X.shape[3])

@d2l.add_to_class(MultiHeadAttention)  #@save
def transpose_output(self, X):
    """Reverse the operation of transpose_qkv."""
    X = X.reshape(-1, self.num_heads, X.shape[1], X.shape[2])
    X = X.permute(0, 2, 1, 3)
    return X.reshape(X.shape[0], X.shape[1], -1)
```

Let's test our implemented `MultiHeadAttention` class using a toy example where keys and values are the same. As a result, the shape of the multi-head attention output is (`batch_size`, `num_queries`, `num_hiddens`).

```
num_hiddens, num_heads = 100, 5
attention = MultiHeadAttention(num_hiddens, num_heads, 0.5)
batch_size, num_queries, num_kvpairs = 2, 4, 6
valid_lens = torch.tensor([3, 2])
X = torch.ones((batch_size, num_queries, num_hiddens))
Y = torch.ones((batch_size, num_kvpairs, num_hiddens))
d2l.check_shape(attention(X, Y, Y, valid_lens),
                (batch_size, num_queries, num_hiddens))
```

## 11.5.3 Summary

Multi-head attention combines knowledge of the same attention pooling via different representation subspaces of queries, keys, and values. To compute multiple heads of multi-head attention in parallel, proper tensor manipulation is needed.

## 11.5.4 Exercises

1. Visualize attention weights of multiple heads in this experiment.
2. Suppose that we have a trained model based on multi-head attention and we want to prune less important attention heads to increase the prediction speed. How can we design experiments to measure the importance of an attention head?

Discussions[160].

160

# 11.6 Self-Attention and Positional Encoding

In deep learning, we often use CNNs or RNNs to encode sequences. Now with attention mechanisms in mind, imagine feeding a sequence of tokens into an attention mechanism such that at every step, each token has its own query, keys, and values. Here, when computing the value of a token's representation at the next layer, the token can attend (via its query vector) to any other's token (matching based on their key vectors). Using the full set of query-key compatibility scores, we can compute, for each token, a representation by building the appropriate weighted sum over the other tokens. Because every token is attending to each other token (unlike the case where decoder steps attend to encoder steps), such architectures are typically described as *self-attention* models (Lin *et al.*, 2017, Vaswani *et al.*, 2017), and elsewhere described as *intra-attention* model (Cheng *et al.*, 2016, Parikh *et al.*, 2016, Paulus *et al.*, 2017). In this section, we will discuss sequence encoding using self-attention, including using additional information for the sequence order.

```
import math
import torch
from torch import nn
from d2l import torch as d2l
```

## 11.6.1 Self-Attention

Given a sequence of input tokens $\mathbf{x}_1, \ldots, \mathbf{x}_n$ where any $\mathbf{x}_i \in \mathbb{R}^d$ $(1 \leq i \leq n)$, its self-attention outputs a sequence of the same length $\mathbf{y}_1, \ldots, \mathbf{y}_n$, where

$$\mathbf{y}_i = f(\mathbf{x}_i, (\mathbf{x}_1, \mathbf{x}_1), \ldots, (\mathbf{x}_n, \mathbf{x}_n)) \in \mathbb{R}^d \tag{11.6.1}$$

according to the definition of attention pooling in (11.1.1). Using multi-head attention, the following code snippet computes the self-attention of a tensor with shape (batch size, number of time steps or sequence length in tokens, $d$). The output tensor has the same shape.

```
num_hiddens, num_heads = 100, 5
attention = d2l.MultiHeadAttention(num_hiddens, num_heads, 0.5)
batch_size, num_queries, valid_lens = 2, 4, torch.tensor([3, 2])
X = torch.ones((batch_size, num_queries, num_hiddens))
d2l.check_shape(attention(X, X, X, valid_lens),
                (batch_size, num_queries, num_hiddens))
```

## 11.6.2 Comparing CNNs, RNNs, and Self-Attention

Let's compare architectures for mapping a sequence of $n$ tokens to another one of equal length, where each input or output token is represented by a $d$-dimensional vector. Specifically, we will consider CNNs, RNNs, and self-attention. We will compare their computational complexity, sequential operations, and maximum path lengths. Note that sequential operations prevent parallel computation, while a shorter path between any combination of sequence positions makes it easier to learn long-range dependencies within the sequence (Hochreiter *et al.*, 2001).

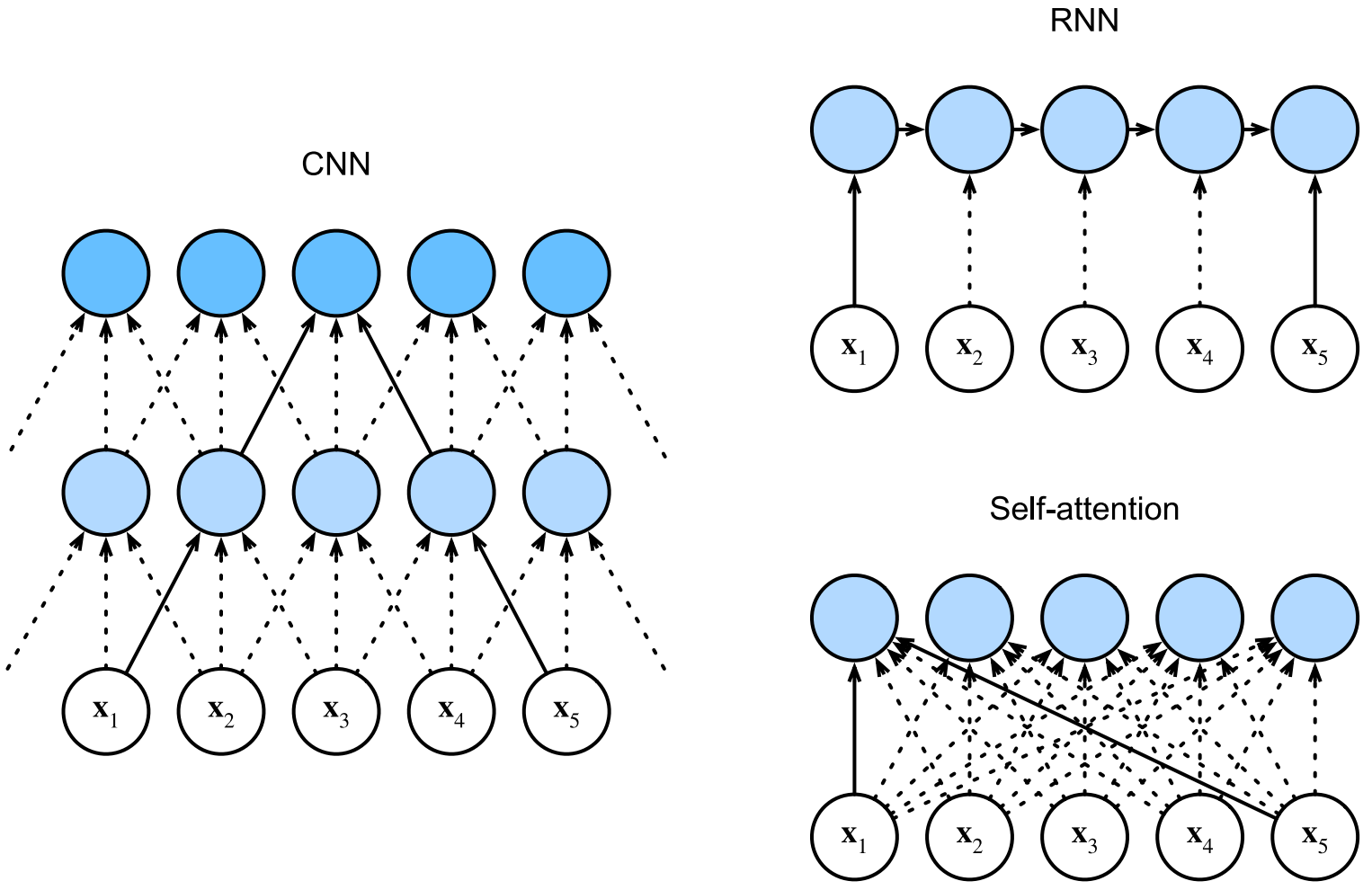

Fig. 11.6.1 Comparing CNN (padding tokens are omitted), RNN, and self-attention architectures.

Let's regard any text sequence as a "one-dimensional image". Similarly, one-dimensional CNNs can process local features such as $n$-grams in text. Given a sequence of length $n$, consider a convolutional layer whose kernel size is $k$, and whose numbers of input and output channels are both $d$. The computational complexity of the convolutional layer is $O(knd^2)$. As Fig. 11.6.1 shows, CNNs are hierarchical, so there are $O(1)$ sequential operations and the maximum path length is $O(n/k)$. For example, $\mathbf{x}_1$ and $\mathbf{x}_5$ are within the receptive field of a two-layer CNN with kernel size 3 in Fig. 11.6.1.

When updating the hidden state of RNNs, multiplication of the $d \times d$ weight matrix and the $d$-dimensional hidden state has a computational complexity of $O(d^2)$. Since the sequence length is $n$, the computational complexity of the recurrent layer is $O(nd^2)$. According to Fig.

11.6.1, there are $O(n)$ sequential operations that cannot be parallelized and the maximum path length is also $O(n)$.

In self-attention, the queries, keys, and values are all $n \times d$ matrices. Consider the scaled dot product attention in (11.3.6), where an $n \times d$ matrix is multiplied by a $d \times n$ matrix, then the output $n \times n$ matrix is multiplied by an $n \times d$ matrix. As a result, the self-attention has a $O(n^2 d)$ computational complexity. As we can see from Fig. 11.6.1, each token is directly connected to any other token via self-attention. Therefore, computation can be parallel with $O(1)$ sequential operations and the maximum path length is also $O(1)$.

All in all, both CNNs and self-attention enjoy parallel computation and self-attention has the shortest maximum path length. However, the quadratic computational complexity with respect to the sequence length makes self-attention prohibitively slow for very long sequences.

## 11.6.3 Positional Encoding

Unlike RNNs, which recurrently process tokens of a sequence one-by-one, self-attention ditches sequential operations in favor of parallel computation. Note that self-attention by itself does not preserve the order of the sequence. What do we do if it really matters that the model knows in which order the input sequence arrived?

The dominant approach for preserving information about the order of tokens is to represent this to the model as an additional input associated with each token. These inputs are called *positional encodings*, and they can either be learned or fixed *a priori*. We now describe a simple scheme for fixed positional encodings based on sine and cosine functions (Vaswani *et al.*, 2017).

Suppose that the input representation $\mathbf{X} \in \mathbb{R}^{n \times d}$ contains the $d$-dimensional embeddings for $n$ tokens of a sequence. The positional encoding outputs $\mathbf{X} + \mathbf{P}$ using a positional embedding matrix $\mathbf{P} \in \mathbb{R}^{n \times d}$ of the same shape, whose element on the $i^\textrm{th}$ row and the $(2j)^\textrm{th}$ or the $(2j+1)^\textrm{th}$ column is

$$\begin{aligned} p_{i, 2j} &= \sin\left(\frac{i}{10000^{2j/d}}\right),\\ p_{i, 2j+1} &= \cos\left(\frac{i}{10000^{2j/d}}\right). \end{aligned} \tag{11.6.2}$$

At first glance, this trigonometric function design looks weird. Before we give explanations of this design, let's first implement it in the following `PositionalEncoding` class.

```
class PositionalEncoding(nn.Module):  #@save
    """Positional encoding."""
    def __init__(self, num_hiddens, dropout, max_len=1000):
        super().__init__()
        self.dropout = nn.Dropout(dropout)
        # Create a long enough P
        self.P = torch.zeros((1, max_len, num_hiddens))
        X = torch.arange(max_len, dtype=torch.float32).reshape(
```

```
            -1, 1) / torch.pow(10000, torch.arange(
            0, num_hiddens, 2, dtype=torch.float32) / num_hiddens)
        self.P[:, :, 0::2] = torch.sin(X)
        self.P[:, :, 1::2] = torch.cos(X)

    def forward(self, X):
        X = X + self.P[:, :X.shape[1], :].to(X.device)
        return self.dropout(X)
```

In the positional embedding matrix $\mathbf{P}$, rows correspond to positions within a sequence and columns represent different positional encoding dimensions. In the example below, we can see that the $6^{\text{th}}$ and the $7^{\text{th}}$ columns of the positional embedding matrix have a higher frequency than the $8^{\text{th}}$ and the $9^{\text{th}}$ columns. The offset between the $6^{\text{th}}$ and the $7^{\text{th}}$ (same for the $8^{\text{th}}$ and the $9^{\text{th}}$) columns is due to the alternation of sine and cosine functions.

```
encoding_dim, num_steps = 32, 60
pos_encoding = PositionalEncoding(encoding_dim, 0)
X = pos_encoding(torch.zeros((1, num_steps, encoding_dim)))
P = pos_encoding.P[:, :X.shape[1], :]
d2l.plot(torch.arange(num_steps), P[0, :, 6:10].T, xlabel='Row (position)',
         figsize=(6, 2.5), legend=["Col %d" % d for d in torch.arange(6, 10)])
```

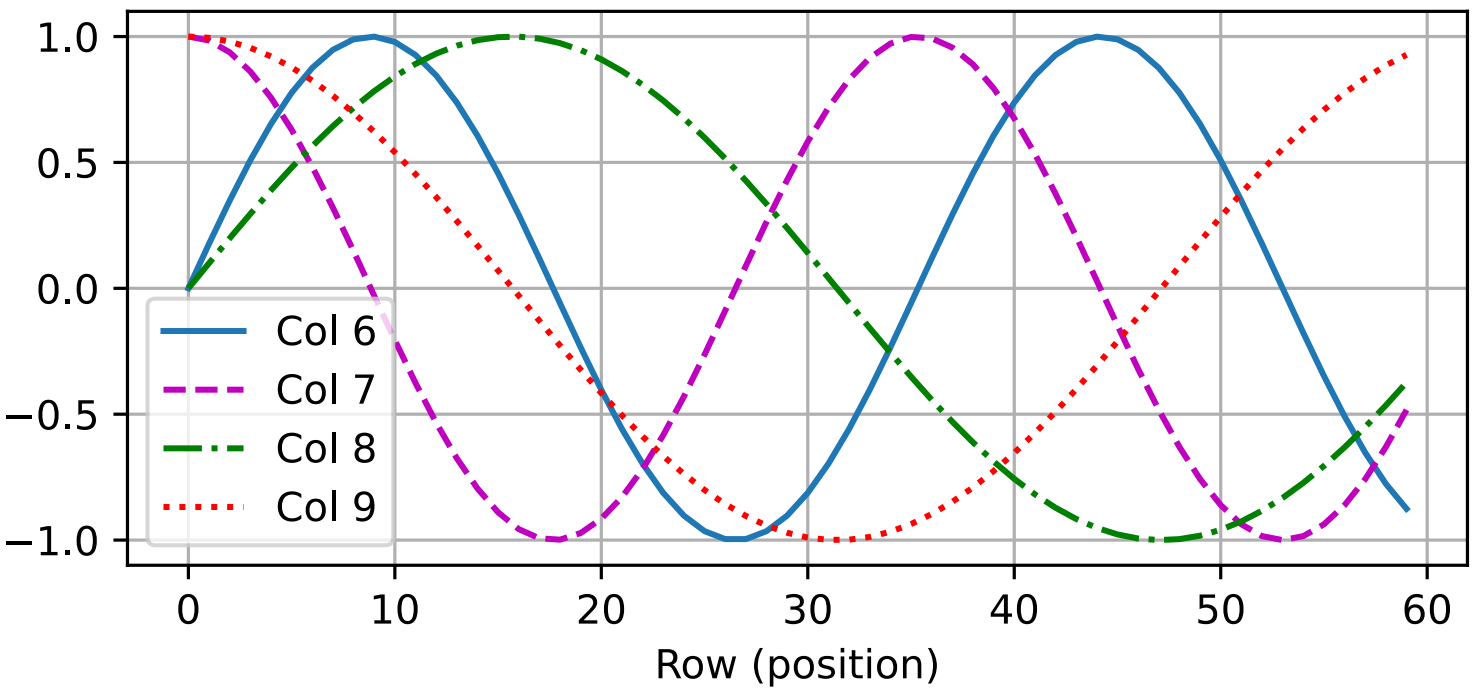

### Absolute Positional Information

To see how the monotonically decreased frequency along the encoding dimension relates to absolute positional information, let's print out the binary representations of $0, 1, \ldots, 7$. As we can see, the lowest bit, the second-lowest bit, and the third-lowest bit alternate on every number, every two numbers, and every four numbers, respectively.

```
for i in range(8):
    print(f'{i} in binary is {i:>03b}')
```

```
0 in binary is 000
1 in binary is 001
2 in binary is 010
3 in binary is 011
4 in binary is 100
5 in binary is 101
6 in binary is 110
7 in binary is 111
```

In binary representations, a higher bit has a lower frequency than a lower bit. Similarly, as demonstrated in the heat map below, the positional encoding decreases frequencies along the encoding dimension by using trigonometric functions. Since the outputs are float numbers, such continuous representations are more space-efficient than binary representations.

```
P = P[0, :, :].unsqueeze(0).unsqueeze(0)
d2l.show_heatmaps(P, xlabel='Column (encoding dimension)',
                  ylabel='Row (position)', figsize=(3.5, 4), cmap='Blues')
```

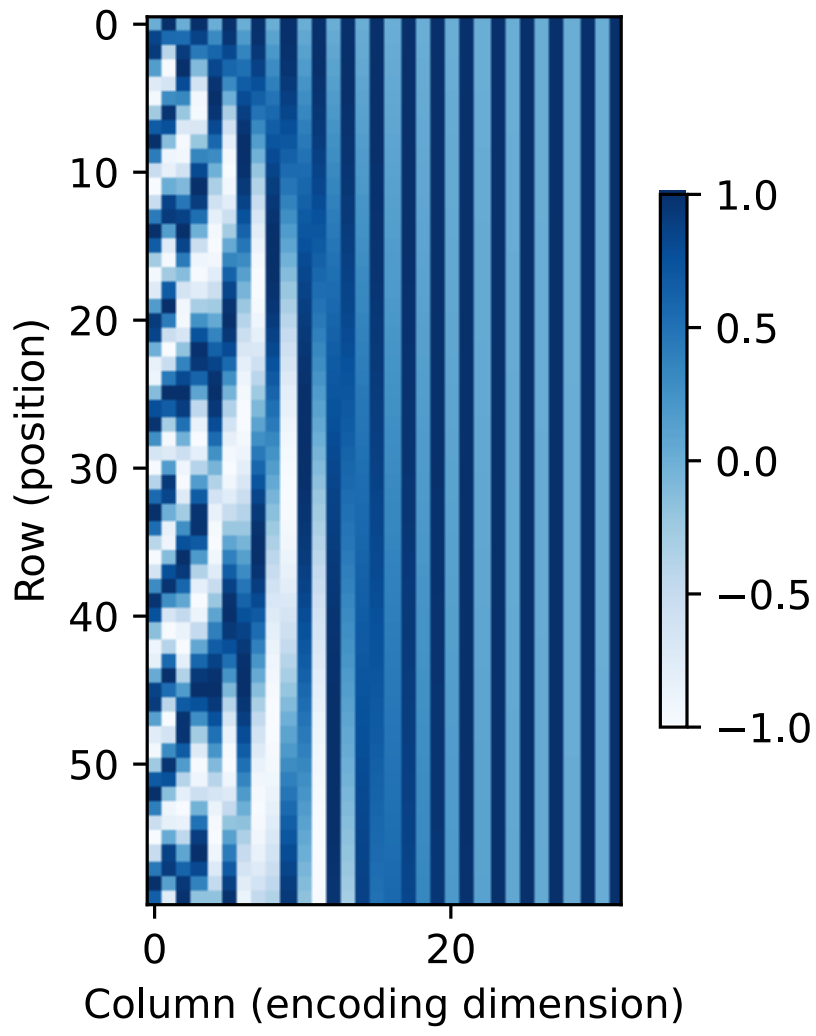

## Relative Positional Information

Besides capturing absolute positional information, the above positional encoding also allows a model to easily learn to attend by relative positions. This is because for any fixed position offset $\delta$, the positional encoding at position $i + \delta$ can be represented by a linear projection of that at position $i$.

This projection can be explained mathematically. Denoting $\omega_j = 1/10000^{2j/d}$, any pair of $(p_{i,2j}, p_{i,2j+1})$ in (11.6.2) can be linearly projected to $(p_{i+\delta,2j}, p_{i+\delta,2j+1})$ for any fixed

offset $\delta$:

$$\begin{aligned}
\begin{bmatrix} \cos(\delta\omega_j) & \sin(\delta\omega_j) \\ -\sin(\delta\omega_j) & \cos(\delta\omega_j) \end{bmatrix}
\begin{bmatrix} p_{i,2j} \\ p_{i,2j+1} \end{bmatrix}
&= \begin{bmatrix} \cos(\delta\omega_j) \sin(i\omega_j) + \sin(\delta\omega_j) \cos(i\omega_j) \\ -\sin(\delta\omega_j) \sin(i\omega_j) + \cos(\delta\omega_j) \cos(i\omega_j) \end{bmatrix} \\
&= \begin{bmatrix} \sin\left((i+\delta)\omega_j\right) \\ \cos\left((i+\delta)\omega_j\right) \end{bmatrix} \\
&= \begin{bmatrix} p_{i+\delta,2j} \\ p_{i+\delta,2j+1} \end{bmatrix},
\end{aligned} \tag{11.6.3}$$

where the $2 \times 2$ projection matrix does not depend on any position index $i$.

### 11.6.4 Summary

In self-attention, the queries, keys, and values all come from the same place. Both CNNs and self-attention enjoy parallel computation and self-attention has the shortest maximum path length. However, the quadratic computational complexity with respect to the sequence length makes self-attention prohibitively slow for very long sequences. To use the sequence order information, we can inject absolute or relative positional information by adding positional encoding to the input representations.

### 11.6.5 Exercises

1. Suppose that we design a deep architecture to represent a sequence by stacking self-attention layers with positional encoding. What could the possible issues be?
2. Can you design a learnable positional encoding method?
3. Can we assign different learned embeddings according to different offsets between queries and keys that are compared in self-attention? Hint: you may refer to relative position embeddings (Huang *et al.*, 2018, Shaw *et al.*, 2018).

Discussions[161].

## 11.7 The Transformer Architecture

We have compared CNNs, RNNs, and self-attention in Section 11.6.2. Notably, self-attention enjoys both parallel computation and the shortest maximum path length. Therefore, it is appealing to design deep architectures by using self-attention. Unlike earlier self-attention models that still rely on RNNs for input representations (Cheng *et al.*, 2016, Lin *et al.*, 2017, Paulus *et al.*, 2017), the Transformer model is solely based on attention mechanisms without any convolutional or recurrent layer (Vaswani *et al.*, 2017). Though originally proposed for sequence-to-sequence learning on text data, Transformers have been pervasive in a wide range of modern deep learning applications, such as in areas to do with language, vision, speech, and reinforcement learning.

```
import math
import pandas as pd
import torch
from torch import nn
from d2l import torch as d2l
```

## 11.7.1 Model

As an instance of the encoder–decoder architecture, the overall architecture of the Transformer is presented in Fig. 11.7.1. As we can see, the Transformer is composed of an encoder and a decoder. In contrast to Bahdanau attention for sequence-to-sequence learning in Fig. 11.4.2, the input (source) and output (target) sequence embeddings are added with positional encoding before being fed into the encoder and the decoder that stack modules based on self-attention.

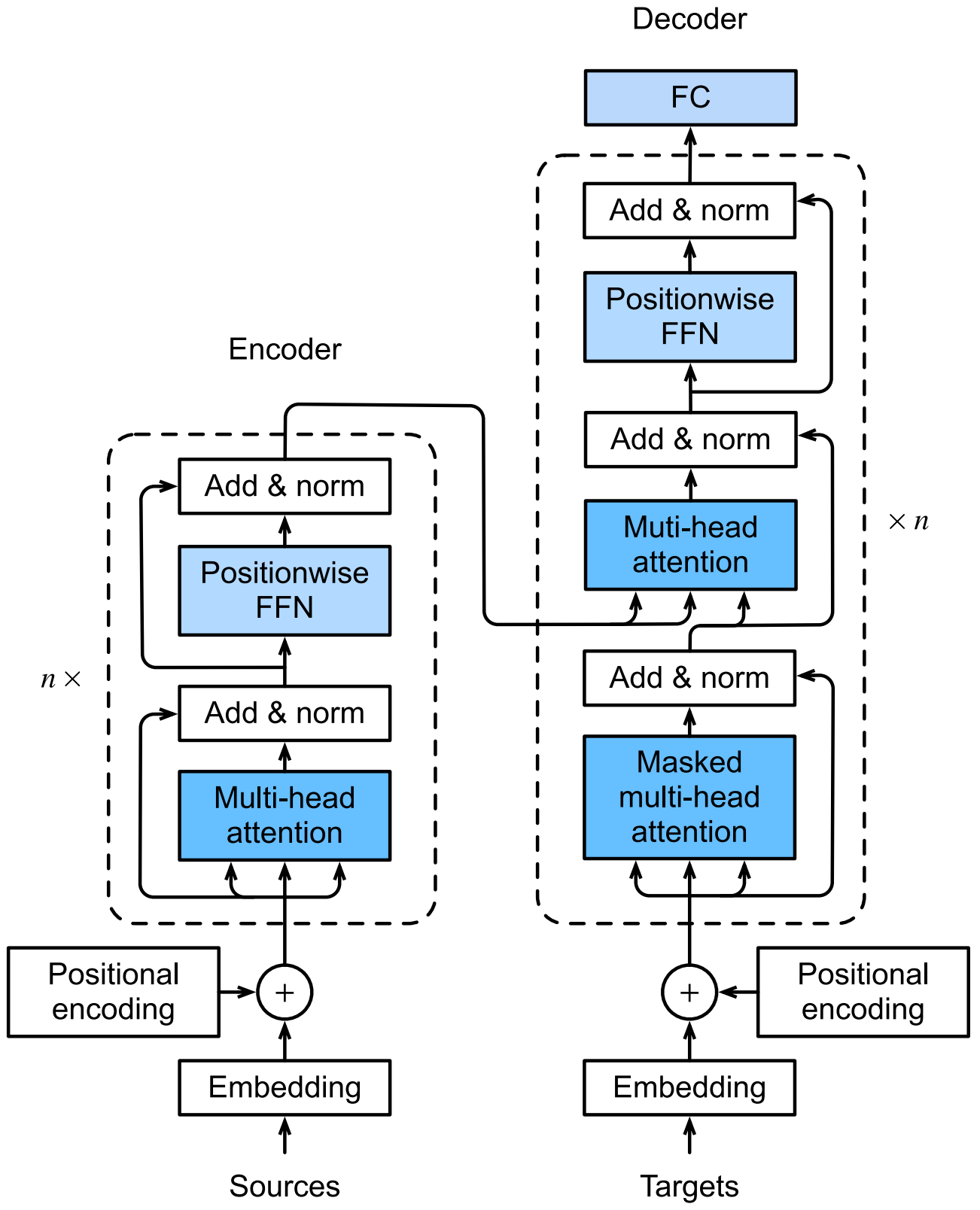

Fig. 11.7.1 The Transformer architecture.

Now we provide an overview of the Transformer architecture in Fig. 11.7.1. At a high level, the Transformer encoder is a stack of multiple identical layers, where each layer has two sublayers (either is denoted as sublayer). The first is a multi-head self-attention pooling and the second is a positionwise feed-forward network. Specifically, in the encoder self-attention, queries, keys, and values are all from the outputs of the previous encoder layer. Inspired by the ResNet design of Section 8.6, a residual connection is employed around both sublayers. In the Transformer, for any input $\mathbf{x} \in \mathbb{R}^d$ at any position of the sequence, we require that $\mathrm{sublayer}(\mathbf{x}) \in \mathbb{R}^d$ so that the residual connection $\mathbf{x} + \mathrm{sublayer}(\mathbf{x}) \in \mathbb{R}^d$ is feasible. This addition from the residual connection is immediately followed by layer normalization (Ba *et al.*, 2016). As a result, the Transformer encoder outputs a $d$-dimensional vector representation for each position of the input sequence.

The Transformer decoder is also a stack of multiple identical layers with residual connections and layer normalizations. As well as the two sublayers described in the encoder, the decoder inserts a third sublayer, known as the encoder–decoder attention, between these two. In the encoder–decoder attention, queries are from the outputs of the decoder's self-attention sublayer, and the keys and values are from the Transformer encoder outputs. In the decoder self-attention, queries, keys, and values are all from the outputs of the previous decoder layer. However, each position in the decoder is allowed only to attend to all positions in the decoder up to that position. This *masked* attention preserves the autoregressive property, ensuring that the prediction only depends on those output tokens that have been generated.

We have already described and implemented multi-head attention based on scaled dot products in Section 11.5 and positional encoding in Section 11.6.3. In the following, we will implement the rest of the Transformer model.

## 11.7.2 Positionwise Feed-Forward Networks

The positionwise feed-forward network transforms the representation at all the sequence positions using the same MLP. This is why we call it *positionwise*. In the implementation below, the input `X` with shape (batch size, number of time steps or sequence length in tokens, number of hidden units or feature dimension) will be transformed by a two-layer MLP into an output tensor of shape (batch size, number of time steps, `ffn_num_outputs`).

```
class PositionWiseFFN(nn.Module):  #@save
    """The positionwise feed-forward network."""
    def __init__(self, ffn_num_hiddens, ffn_num_outputs):
        super().__init__()
        self.dense1 = nn.LazyLinear(ffn_num_hiddens)
        self.relu = nn.ReLU()
        self.dense2 = nn.LazyLinear(ffn_num_outputs)

    def forward(self, X):
        return self.dense2(self.relu(self.dense1(X)))
```

The following example shows that the innermost dimension of a tensor changes to the number of outputs in the positionwise feed-forward network. Since the same MLP transforms at all the positions, when the inputs at all these positions are the same, their outputs are also identical.

```
ffn = PositionWiseFFN(4, 8)
ffn.eval()
ffn(torch.ones((2, 3, 4)))[0]
```

```
tensor([[-0.0458,  0.2515, -0.4782, -0.0908, -0.3782,  0.4959, -0.1183, -0.
↪0020],
        [-0.0458,  0.2515, -0.4782, -0.0908, -0.3782,  0.4959, -0.1183, -0.
↪0020],
        [-0.0458,  0.2515, -0.4782, -0.0908, -0.3782,  0.4959, -0.1183, -0.
↪0020]],
       grad_fn=<SelectBackward0>)
```

### 11.7.3 Residual Connection and Layer Normalization

Now let's focus on the "add & norm" component in Fig. 11.7.1. As we described at the beginning of this section, this is a residual connection immediately followed by layer normalization. Both are key to effective deep architectures.

In Section 8.5, we explained how batch normalization recenters and rescales across the examples within a minibatch. As discussed in Section 8.5.2, layer normalization is the same as batch normalization except that the former normalizes across the feature dimension, thus enjoying benefits of scale independence and batch size independence. Despite its pervasive applications in computer vision, batch normalization is usually empirically less effective than layer normalization in natural language processing tasks, where the inputs are often variable-length sequences.

The following code snippet compares the normalization across different dimensions by layer normalization and batch normalization.

```
ln = nn.LayerNorm(2)
bn = nn.LazyBatchNorm1d()
X = torch.tensor([[1, 2], [2, 3]], dtype=torch.float32)
# Compute mean and variance from X in the training mode
print('layer norm:', ln(X), '\nbatch norm:', bn(X))
```

```
layer norm: tensor([[-1.0000,  1.0000],
        [-1.0000,  1.0000]], grad_fn=<NativeLayerNormBackward0>)
batch norm: tensor([[-1.0000, -1.0000],
        [ 1.0000,  1.0000]], grad_fn=<NativeBatchNormBackward0>)
```

Now we can implement the `AddNorm` class using a residual connection followed by layer normalization. Dropout is also applied for regularization.

```
class AddNorm(nn.Module):  #@save
    """The residual connection followed by layer normalization."""
    def __init__(self, norm_shape, dropout):
        super().__init__()
        self.dropout = nn.Dropout(dropout)
        self.ln = nn.LayerNorm(norm_shape)

    def forward(self, X, Y):
        return self.ln(self.dropout(Y) + X)
```

The residual connection requires that the two inputs are of the same shape so that the output tensor also has the same shape after the addition operation.

```
add_norm = AddNorm(4, 0.5)
shape = (2, 3, 4)
d2l.check_shape(add_norm(torch.ones(shape), torch.ones(shape)), shape)
```

### 11.7.4 Encoder

With all the essential components to assemble the Transformer encoder, let's start by implementing a single layer within the encoder. The following `TransformerEncoderBlock` class contains two sublayers: multi-head self-attention and positionwise feed-forward networks, where a residual connection followed by layer normalization is employed around both sublayers.

```
class TransformerEncoderBlock(nn.Module):  #@save
    """The Transformer encoder block."""
    def __init__(self, num_hiddens, ffn_num_hiddens, num_heads, dropout,
                 use_bias=False):
        super().__init__()
        self.attention = d2l.MultiHeadAttention(num_hiddens, num_heads,
                                                dropout, use_bias)
        self.addnorm1 = AddNorm(num_hiddens, dropout)
        self.ffn = PositionWiseFFN(ffn_num_hiddens, num_hiddens)
        self.addnorm2 = AddNorm(num_hiddens, dropout)

    def forward(self, X, valid_lens):
        Y = self.addnorm1(X, self.attention(X, X, X, valid_lens))
        return self.addnorm2(Y, self.ffn(Y))
```

As we can see, no layer in the Transformer encoder changes the shape of its input.

```
X = torch.ones((2, 100, 24))
valid_lens = torch.tensor([3, 2])
encoder_blk = TransformerEncoderBlock(24, 48, 8, 0.5)
encoder_blk.eval()
d2l.check_shape(encoder_blk(X, valid_lens), X.shape)
```

In the following Transformer encoder implementation, we stack `num_blks` instances of the above `TransformerEncoderBlock` classes. Since we use the fixed positional encoding whose values are always between $-1$ and $1$, we multiply values of the learnable input embeddings by the square root of the embedding dimension to rescale before summing up the input embedding and the positional encoding.

```
class TransformerEncoder(d2l.Encoder):  #@save
    """The Transformer encoder."""
    def __init__(self, vocab_size, num_hiddens, ffn_num_hiddens,
                 num_heads, num_blks, dropout, use_bias=False):
        super().__init__()
        self.num_hiddens = num_hiddens
        self.embedding = nn.Embedding(vocab_size, num_hiddens)
        self.pos_encoding = d2l.PositionalEncoding(num_hiddens, dropout)
        self.blks = nn.Sequential()
        for i in range(num_blks):
            self.blks.add_module("block"+str(i), TransformerEncoderBlock(
                num_hiddens, ffn_num_hiddens, num_heads, dropout, use_bias))

    def forward(self, X, valid_lens):
        # Since positional encoding values are between -1 and 1, the embedding
        # values are multiplied by the square root of the embedding dimension
        # to rescale before they are summed up
        X = self.pos_encoding(self.embedding(X) * math.sqrt(self.num_hiddens))
        self.attention_weights = [None] * len(self.blks)
        for i, blk in enumerate(self.blks):
            X = blk(X, valid_lens)
            self.attention_weights[
                i] = blk.attention.attention.attention_weights
        return X
```

Below we specify hyperparameters to create a two-layer Transformer encoder. The shape of the Transformer encoder output is (batch size, number of time steps, `num_hiddens`).

```
encoder = TransformerEncoder(200, 24, 48, 8, 2, 0.5)
d2l.check_shape(encoder(torch.ones((2, 100), dtype=torch.long), valid_lens),
                (2, 100, 24))
```

## 11.7.5 Decoder

As shown in Fig. 11.7.1, the Transformer decoder is composed of multiple identical layers. Each layer is implemented in the following `TransformerDecoderBlock` class, which contains three sublayers: decoder self-attention, encoder–decoder attention, and positionwise feed-forward networks. These sublayers employ a residual connection around them followed by layer normalization.

As we described earlier in this section, in the masked multi-head decoder self-attention (the first sublayer), queries, keys, and values all come from the outputs of the previous decoder layer. When training sequence-to-sequence models, tokens at all the positions (time steps) of the output sequence are known. However, during prediction the output sequence is generated token by token; thus, at any decoder time step only the generated tokens can be used in the decoder self-attention. To preserve autoregression in the decoder, its masked self-attention specifies `dec_valid_lens` so that any query only attends to all positions in the decoder up to the query position.

```
class TransformerDecoderBlock(nn.Module):
    # The i-th block in the Transformer decoder
    def __init__(self, num_hiddens, ffn_num_hiddens, num_heads, dropout, i):
        super().__init__()
        self.i = i
        self.attention1 = d2l.MultiHeadAttention(num_hiddens, num_heads,
                                                 dropout)
        self.addnorm1 = AddNorm(num_hiddens, dropout)
        self.attention2 = d2l.MultiHeadAttention(num_hiddens, num_heads,
                                                 dropout)
        self.addnorm2 = AddNorm(num_hiddens, dropout)
        self.ffn = PositionWiseFFN(ffn_num_hiddens, num_hiddens)
        self.addnorm3 = AddNorm(num_hiddens, dropout)

    def forward(self, X, state):
        enc_outputs, enc_valid_lens = state[0], state[1]
        # During training, all the tokens of any output sequence are processed
        # at the same time, so state[2][self.i] is None as initialized. When
        # decoding any output sequence token by token during prediction,
        # state[2][self.i] contains representations of the decoded output at
        # the i-th block up to the current time step
        if state[2][self.i] is None:
            key_values = X
        else:
            key_values = torch.cat((state[2][self.i], X), dim=1)
        state[2][self.i] = key_values
        if self.training:
            batch_size, num_steps, _ = X.shape
            # Shape of dec_valid_lens: (batch_size, num_steps), where every
            # row is [1, 2, ..., num_steps]
            dec_valid_lens = torch.arange(
                1, num_steps + 1, device=X.device).repeat(batch_size, 1)
        else:
            dec_valid_lens = None
        # Self-attention
        X2 = self.attention1(X, key_values, key_values, dec_valid_lens)
        Y = self.addnorm1(X, X2)
        # Encoder-decoder attention. Shape of enc_outputs:
        # (batch_size, num_steps, num_hiddens)
        Y2 = self.attention2(Y, enc_outputs, enc_outputs, enc_valid_lens)
        Z = self.addnorm2(Y, Y2)
        return self.addnorm3(Z, self.ffn(Z)), state
```

To facilitate scaled dot product operations in the encoder–decoder attention and addition operations in the residual connections, the feature dimension (`num_hiddens`) of the decoder is the same as that of the encoder.

```
decoder_blk = TransformerDecoderBlock(24, 48, 8, 0.5, 0)
X = torch.ones((2, 100, 24))
```

```
state = [encoder_blk(X, valid_lens), valid_lens, [None]]
d2l.check_shape(decoder_blk(X, state)[0], X.shape)
```

Now we construct the entire Transformer decoder composed of `num_blks` instances of `TransformerDecoderBlock`. In the end, a fully connected layer computes the prediction for all the `vocab_size` possible output tokens. Both of the decoder self-attention weights and the encoder–decoder attention weights are stored for later visualization.

```
class TransformerDecoder(d2l.AttentionDecoder):
    def __init__(self, vocab_size, num_hiddens, ffn_num_hiddens, num_heads,
                 num_blks, dropout):
        super().__init__()
        self.num_hiddens = num_hiddens
        self.num_blks = num_blks
        self.embedding = nn.Embedding(vocab_size, num_hiddens)
        self.pos_encoding = d2l.PositionalEncoding(num_hiddens, dropout)
        self.blks = nn.Sequential()
        for i in range(num_blks):
            self.blks.add_module("block"+str(i), TransformerDecoderBlock(
                num_hiddens, ffn_num_hiddens, num_heads, dropout, i))
        self.dense = nn.LazyLinear(vocab_size)

    def init_state(self, enc_outputs, enc_valid_lens):
        return [enc_outputs, enc_valid_lens, [None] * self.num_blks]

    def forward(self, X, state):
        X = self.pos_encoding(self.embedding(X) * math.sqrt(self.num_hiddens))
        self._attention_weights = [[None] * len(self.blks) for _ in range (2)]
        for i, blk in enumerate(self.blks):
            X, state = blk(X, state)
            # Decoder self-attention weights
            self._attention_weights[0][
                i] = blk.attention1.attention.attention_weights
            # Encoder-decoder attention weights
            self._attention_weights[1][
                i] = blk.attention2.attention.attention_weights
        return self.dense(X), state

    @property
    def attention_weights(self):
        return self._attention_weights
```

### 11.7.6 Training

Let's instantiate an encoder–decoder model by following the Transformer architecture. Here we specify that both the Transformer encoder and the Transformer decoder have two layers using 4-head attention. As in Section 10.7.6, we train the Transformer model for sequence-to-sequence learning on the English–French machine translation dataset.

```
data = d2l.MTFraEng(batch_size=128)
num_hiddens, num_blks, dropout = 256, 2, 0.2
ffn_num_hiddens, num_heads = 64, 4
encoder = TransformerEncoder(
    len(data.src_vocab), num_hiddens, ffn_num_hiddens, num_heads,
    num_blks, dropout)
decoder = TransformerDecoder(
    len(data.tgt_vocab), num_hiddens, ffn_num_hiddens, num_heads,
    num_blks, dropout)
model = d2l.Seq2Seq(encoder, decoder, tgt_pad=data.tgt_vocab['<pad>'],
                    lr=0.001)
trainer = d2l.Trainer(max_epochs=30, gradient_clip_val=1, num_gpus=1)
trainer.fit(model, data)
```

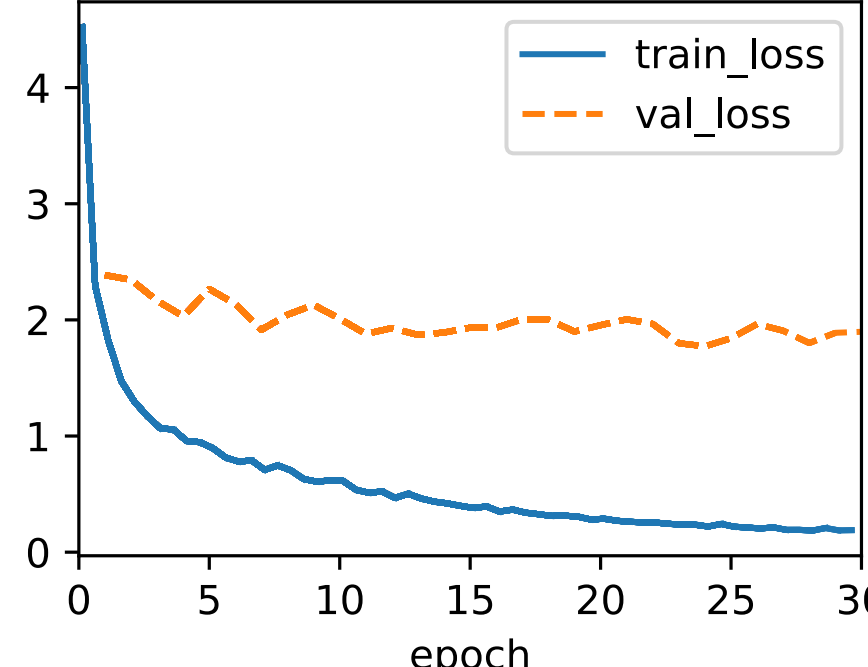

After training, we use the Transformer model to translate a few English sentences into French and compute their BLEU scores.

```
engs = ['go .', 'i lost .', 'he\'s calm .', 'i\'m home .']
fras = ['va !', 'j\'ai perdu .', 'il est calme .', 'je suis chez moi .']
preds, _ = model.predict_step(
    data.build(engs, fras), d2l.try_gpu(), data.num_steps)
for en, fr, p in zip(engs, fras, preds):
    translation = []
    for token in data.tgt_vocab.to_tokens(p):
        if token == '<eos>':
            break
        translation.append(token)
    print(f'{en} => {translation}, bleu,'
          f'{d2l.bleu(" ".join(translation), fr, k=2):.3f}')
```

```
go . => ['va', '!'], bleu,1.000
i lost . => ["j'ai", 'perdu', '.'], bleu,1.000
he's calm . => ['il', 'est', 'mouillé', '.'], bleu,0.658
i'm home . => ['je', 'suis', 'chez', 'moi', '.'], bleu,1.000
```

Let's visualize the Transformer attention weights when translating the final English sentence into French. The shape of the encoder self-attention weights is (number of encoder layers, number of attention heads, num_steps or number of queries, num_steps or number of key-value pairs).

```
_, dec_attention_weights = model.predict_step(
    data.build([engs[-1]], [fras[-1]]), d2l.try_gpu(), data.num_steps, True)
enc_attention_weights = torch.cat(model.encoder.attention_weights, 0)
shape = (num_blks, num_heads, -1, data.num_steps)
enc_attention_weights = enc_attention_weights.reshape(shape)
d2l.check_shape(enc_attention_weights,
                (num_blks, num_heads, data.num_steps, data.num_steps))
```

In the encoder self-attention, both queries and keys come from the same input sequence. Since padding tokens do not carry meaning, with specified valid length of the input sequence no query attends to positions of padding tokens. In the following, two layers of multi-head attention weights are presented row by row. Each head independently attends based on a separate representation subspace of queries, keys, and values.

```
d2l.show_heatmaps(
    enc_attention_weights.cpu(), xlabel='Key positions',
    ylabel='Query positions', titles=['Head %d' % i for i in range(1, 5)],
    figsize=(7, 3.5))
```

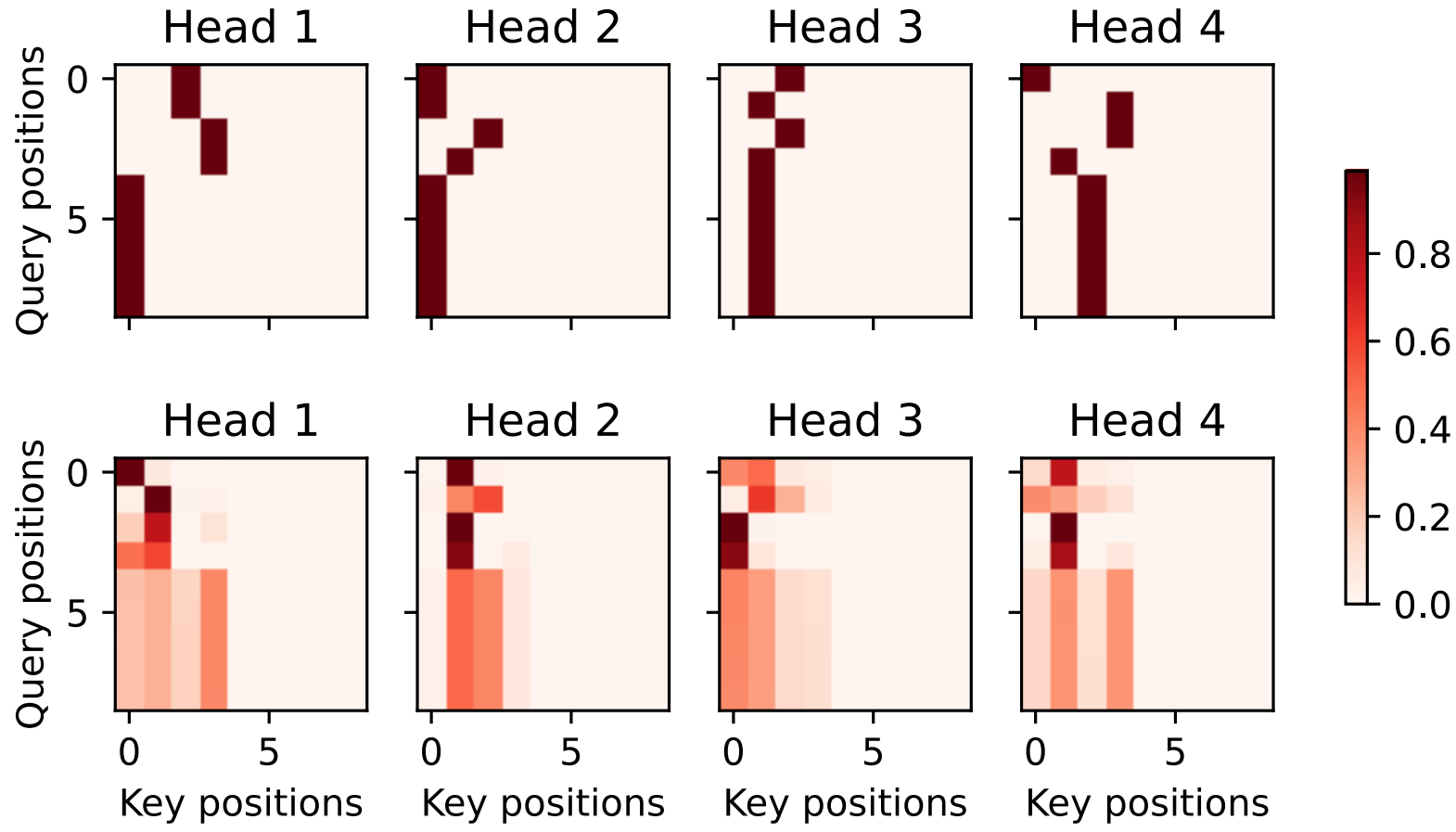

To visualize the decoder self-attention weights and the encoder–decoder attention weights, we need more data manipulations. For example, we fill the masked attention weights with zero. Note that the decoder self-attention weights and the encoder–decoder attention weights both have the same queries: the beginning-of-sequence token followed by the output tokens and possibly end-of-sequence tokens.

```
dec_attention_weights_2d = [head[0].tolist()
                            for step in dec_attention_weights
                            for attn in step for blk in attn for head in blk]
dec_attention_weights_filled = torch.tensor(
```

(continued from previous page)

```
    pd.DataFrame(dec_attention_weights_2d).fillna(0.0).values)
shape = (-1, 2, num_blks, num_heads, data.num_steps)
dec_attention_weights = dec_attention_weights_filled.reshape(shape)
dec_self_attention_weights, dec_inter_attention_weights = \
    dec_attention_weights.permute(1, 2, 3, 0, 4)
```

```
d2l.check_shape(dec_self_attention_weights,
                (num_blks, num_heads, data.num_steps, data.num_steps))
d2l.check_shape(dec_inter_attention_weights,
                (num_blks, num_heads, data.num_steps, data.num_steps))
```

Because of the autoregressive property of the decoder self-attention, no query attends to key–value pairs after the query position.

```
d2l.show_heatmaps(
    dec_self_attention_weights[:, :, :, :],
    xlabel='Key positions', ylabel='Query positions',
    titles=['Head %d' % i for i in range(1, 5)], figsize=(7, 3.5))
```

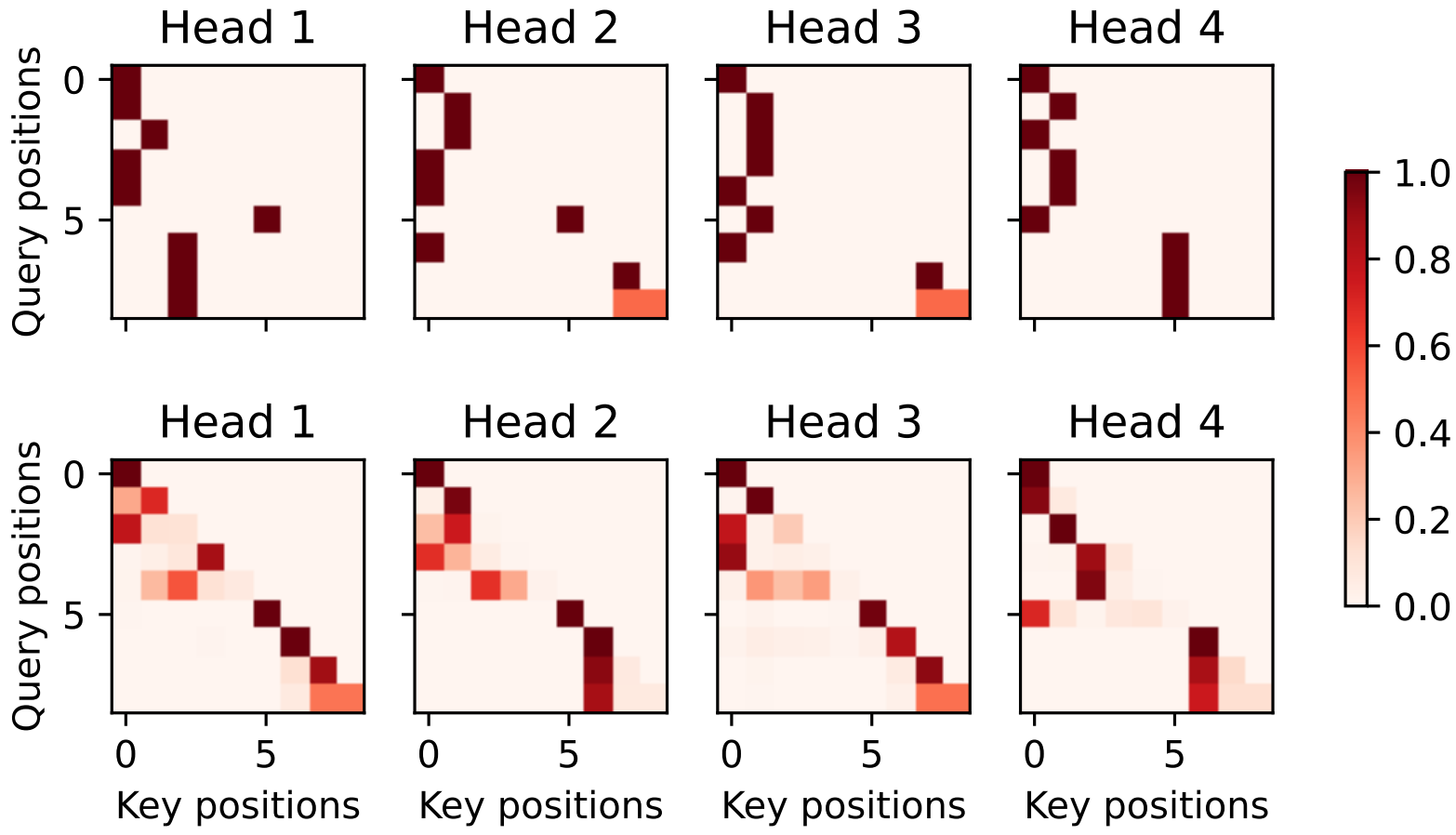

Similar to the case in the encoder self-attention, via the specified valid length of the input sequence, no query from the output sequence attends to those padding tokens from the input sequence.

```
d2l.show_heatmaps(
    dec_inter_attention_weights, xlabel='Key positions',
    ylabel='Query positions', titles=['Head %d' % i for i in range(1, 5)],
    figsize=(7, 3.5))
```

Although the Transformer architecture was originally proposed for sequence-to-sequence learning, as we will discover later in the book, either the Transformer encoder or the Transformer decoder is often individually used for different deep learning tasks.

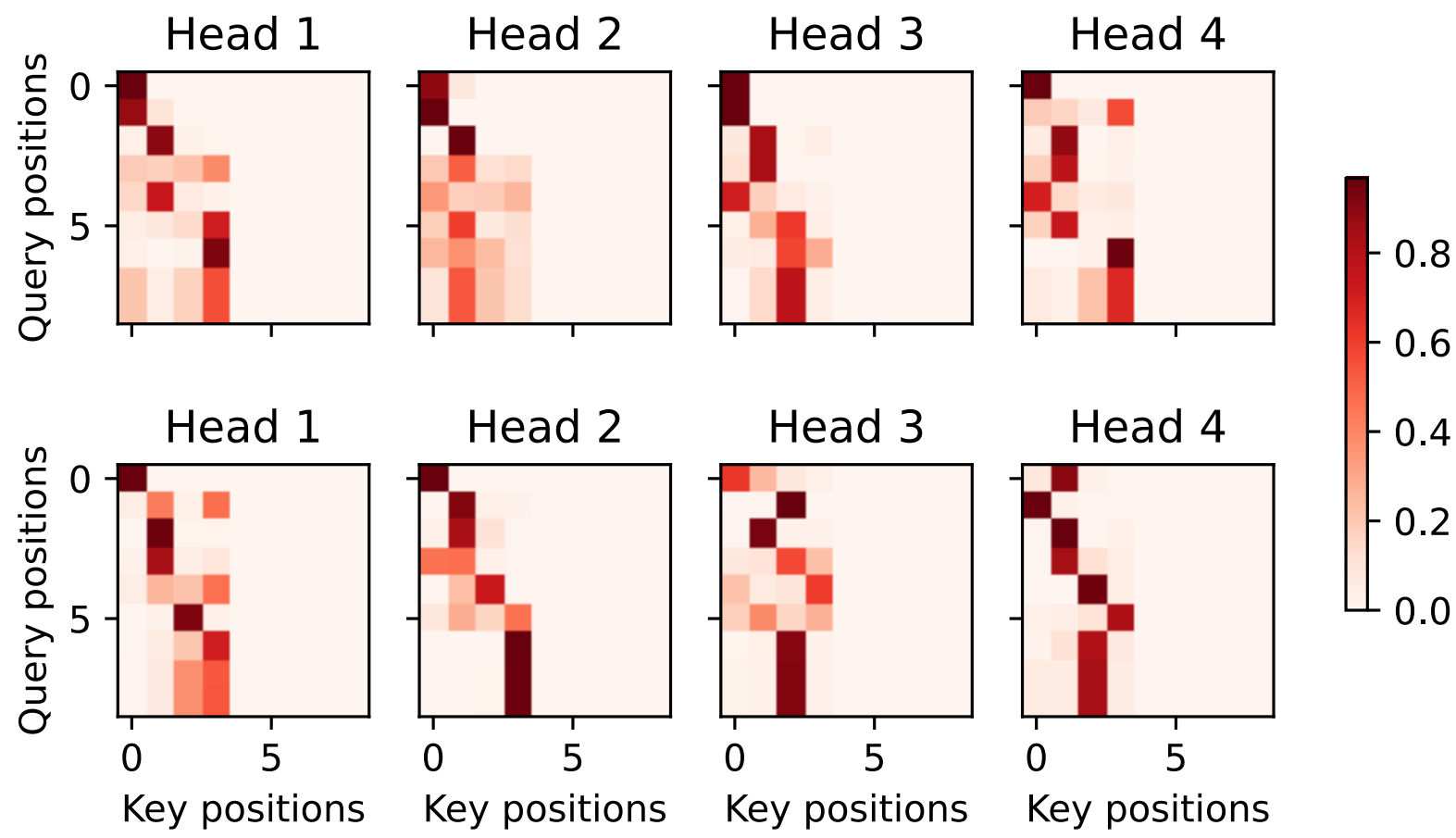

## 11.7.7 Summary

The Transformer is an instance of the encoder–decoder architecture, though either the encoder or the decoder can be used individually in practice. In the Transformer architecture, multi-head self-attention is used for representing the input sequence and the output sequence, though the decoder has to preserve the autoregressive property via a masked version. Both the residual connections and the layer normalization in the Transformer are important for training a very deep model. The positionwise feed-forward network in the Transformer model transforms the representation at all the sequence positions using the same MLP.

## 11.7.8 Exercises

1. Train a deeper Transformer in the experiments. How does it affect the training speed and the translation performance?
2. Is it a good idea to replace scaled dot product attention with additive attention in the Transformer? Why?
3. For language modeling, should we use the Transformer encoder, decoder, or both? How would you design this method?
4. What challenges can Transformers face if input sequences are very long? Why?
5. How would you improve the computational and memory efficiency of Transformers? Hint: you may refer to the survey paper by Tay *et al.* (2020).

Discussions[162].

# 11.8 Transformers for Vision

The Transformer architecture was initially proposed for sequence-to-sequence learning, with a focus on machine translation. Subsequently, Transformers emerged as the model of choice in various natural language processing tasks (Brown *et al.*, 2020, Devlin *et al.*, 2018, Radford *et al.*, 2018, Radford *et al.*, 2019, Raffel *et al.*, 2020). However, in the field of computer vision the dominant architecture has remained the CNN (Chapter 8). Naturally, researchers started to wonder if it might be possible to do better by adapting Transformer models to image data. This question sparked immense interest in the computer vision community. Recently, Ramachandran *et al.* (2019) proposed a scheme for replacing convolution with self-attention. However, its use of specialized patterns in attention makes it hard to scale up models on hardware accelerators. Then, Cordonnier *et al.* (2020) theoretically proved that self-attention can learn to behave similarly to convolution. Empirically, $2 \times 2$ patches were taken from images as inputs, but the small patch size makes the model only applicable to image data with low resolutions.

Without specific constraints on patch size, *vision Transformers* (ViTs) extract patches from images and feed them into a Transformer encoder to obtain a global representation, which will finally be transformed for classification (Dosovitskiy *et al.*, 2021). Notably, Transformers show better scalability than CNNs: and when training larger models on larger datasets, vision Transformers outperform ResNets by a significant margin. Similar to the landscape of network architecture design in natural language processing, Transformers have also become a game-changer in computer vision.

```
import torch
from torch import nn
from d2l import torch as d2l
```

## 11.8.1 Model

Fig. 11.8.1 depicts the model architecture of vision Transformers. This architecture consists of a stem that patchifies images, a body based on the multilayer Transformer encoder, and a head that transforms the global representation into the output label.

Consider an input image with height $h$, width $w$, and $c$ channels. Specifying the patch height and width both as $p$, the image is split into a sequence of $m = hw/p^2$ patches, where each patch is flattened to a vector of length $cp^2$. In this way, image patches can be treated similarly to tokens in text sequences by Transformer encoders. A special "<cls>" (class) token and the $m$ flattened image patches are linearly projected into a sequence of $m + 1$ vectors, summed with learnable positional embeddings. The multilayer Transformer encoder transforms $m + 1$ input vectors into the same number of output vector representations of the same length. It works exactly the same way as the original Transformer encoder in Fig. 11.7.1, only differing in the position of normalization. Since the "<cls>" token attends to all the image patches via self-attention (see Fig. 11.6.1), its representation from the Transformer encoder output will be further transformed into the output label.

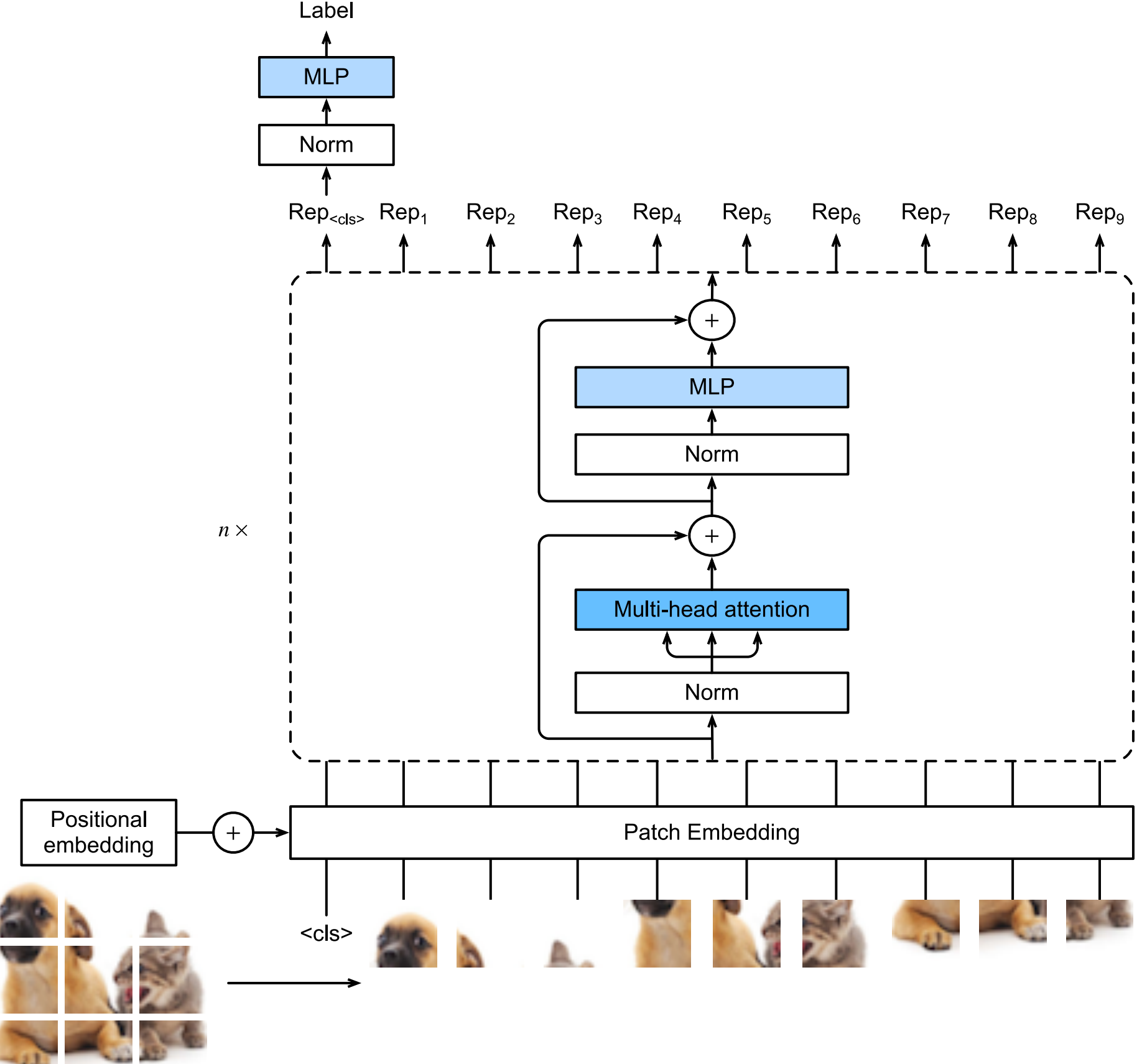

Fig. 11.8.1 The vision Transformer architecture. In this example, an image is split into nine patches. A special “<cls>” token and the nine flattened image patches are transformed via patch embedding and $n$ Transformer encoder blocks into ten representations, respectively. The “<cls>” representation is further transformed into the output label.

## 11.8.2 Patch Embedding

To implement a vision Transformer, let’s start with patch embedding in Fig. 11.8.1. Splitting an image into patches and linearly projecting these flattened patches can be simplified as a single convolution operation, where both the kernel size and the stride size are set to the patch size.

```
class PatchEmbedding(nn.Module):
    def __init__(self, img_size=96, patch_size=16, num_hiddens=512):
```

```
        super().__init__()
        def _make_tuple(x):
            if not isinstance(x, (list, tuple)):
                return (x, x)
            return x
        img_size, patch_size = _make_tuple(img_size), _make_tuple(patch_size)
        self.num_patches = (img_size[0] // patch_size[0]) * (
            img_size[1] // patch_size[1])
        self.conv = nn.LazyConv2d(num_hiddens, kernel_size=patch_size,
                                  stride=patch_size)

    def forward(self, X):
        # Output shape: (batch size, no. of patches, no. of channels)
        return self.conv(X).flatten(2).transpose(1, 2)
```

In the following example, taking images with height and width of `img_size` as inputs, the patch embedding outputs `(img_size//patch_size)**2` patches that are linearly projected to vectors of length `num_hiddens`.

```
img_size, patch_size, num_hiddens, batch_size = 96, 16, 512, 4
patch_emb = PatchEmbedding(img_size, patch_size, num_hiddens)
X = torch.zeros(batch_size, 3, img_size, img_size)
d2l.check_shape(patch_emb(X),
                (batch_size, (img_size//patch_size)**2, num_hiddens))
```

### 11.8.3 Vision Transformer Encoder

The MLP of the vision Transformer encoder is slightly different from the positionwise FFN of the original Transformer encoder (see Section 11.7.2). First, here the activation function uses the Gaussian error linear unit (GELU), which can be considered as a smoother version of the ReLU (Hendrycks and Gimpel, 2016). Second, dropout is applied to the output of each fully connected layer in the MLP for regularization.

```
class ViTMLP(nn.Module):
    def __init__(self, mlp_num_hiddens, mlp_num_outputs, dropout=0.5):
        super().__init__()
        self.dense1 = nn.LazyLinear(mlp_num_hiddens)
        self.gelu = nn.GELU()
        self.dropout1 = nn.Dropout(dropout)
        self.dense2 = nn.LazyLinear(mlp_num_outputs)
        self.dropout2 = nn.Dropout(dropout)

    def forward(self, x):
        return self.dropout2(self.dense2(self.dropout1(self.gelu(
            self.dense1(x)))))
```

The vision Transformer encoder block implementation just follows the pre-normalization design in Fig. 11.8.1, where normalization is applied right *before* multi-head attention or the MLP. In contrast to post-normalization ("add & norm" in Fig. 11.7.1), where normalization is placed right *after* residual connections, pre-normalization leads to more effective or efficient training for Transformers (Baevski and Auli, 2018, Wang *et al.*, 2019, Xiong *et al.*, 2020).

```
class ViTBlock(nn.Module):
    def __init__(self, num_hiddens, norm_shape, mlp_num_hiddens,
                 num_heads, dropout, use_bias=False):
        super().__init__()
        self.ln1 = nn.LayerNorm(norm_shape)
        self.attention = d2l.MultiHeadAttention(num_hiddens, num_heads,
                                                dropout, use_bias)
        self.ln2 = nn.LayerNorm(norm_shape)
        self.mlp = ViTMLP(mlp_num_hiddens, num_hiddens, dropout)

    def forward(self, X, valid_lens=None):
        X = X + self.attention(*([self.ln1(X)] * 3), valid_lens)
        return X + self.mlp(self.ln2(X))
```

Just as in Section 11.7.4, no vision Transformer encoder block changes its input shape.

```
X = torch.ones((2, 100, 24))
encoder_blk = ViTBlock(24, 24, 48, 8, 0.5)
encoder_blk.eval()
d2l.check_shape(encoder_blk(X), X.shape)
```

### 11.8.4 Putting It All Together

The forward pass of vision Transformers below is straightforward. First, input images are fed into an `PatchEmbedding` instance, whose output is concatenated with the “<cls>” token embedding. They are summed with learnable positional embeddings before dropout. Then the output is fed into the Transformer encoder that stacks `num_blks` instances of the `ViTBlock` class. Finally, the representation of the “<cls>” token is projected by the network head.

```
class ViT(d2l.Classifier):
    """Vision Transformer."""
    def __init__(self, img_size, patch_size, num_hiddens, mlp_num_hiddens,
                 num_heads, num_blks, emb_dropout, blk_dropout, lr=0.1,
                 use_bias=False, num_classes=10):
        super().__init__()
        self.save_hyperparameters()
        self.patch_embedding = PatchEmbedding(
            img_size, patch_size, num_hiddens)
        self.cls_token = nn.Parameter(torch.zeros(1, 1, num_hiddens))
        num_steps = self.patch_embedding.num_patches + 1  # Add the cls token
        # Positional embeddings are learnable
        self.pos_embedding = nn.Parameter(
            torch.randn(1, num_steps, num_hiddens))
        self.dropout = nn.Dropout(emb_dropout)
        self.blks = nn.Sequential()
```

```
        for i in range(num_blks):
            self.blks.add_module(f"{i}", ViTBlock(
                num_hiddens, num_hiddens, mlp_num_hiddens,
                num_heads, blk_dropout, use_bias))
        self.head = nn.Sequential(nn.LayerNorm(num_hiddens),
                                  nn.Linear(num_hiddens, num_classes))

    def forward(self, X):
        X = self.patch_embedding(X)
        X = torch.cat((self.cls_token.expand(X.shape[0], -1, -1), X), 1)
        X = self.dropout(X + self.pos_embedding)
        for blk in self.blks:
            X = blk(X)
        return self.head(X[:, 0])
```

## 11.8.5 Training

Training a vision Transformer on the Fashion-MNIST dataset is just like how CNNs were trained in Chapter 8.

```
img_size, patch_size = 96, 16
num_hiddens, mlp_num_hiddens, num_heads, num_blks = 512, 2048, 8, 2
emb_dropout, blk_dropout, lr = 0.1, 0.1, 0.1
model = ViT(img_size, patch_size, num_hiddens, mlp_num_hiddens, num_heads,
            num_blks, emb_dropout, blk_dropout, lr)
trainer = d2l.Trainer(max_epochs=10, num_gpus=1)
data = d2l.FashionMNIST(batch_size=128, resize=(img_size, img_size))
trainer.fit(model, data)
```

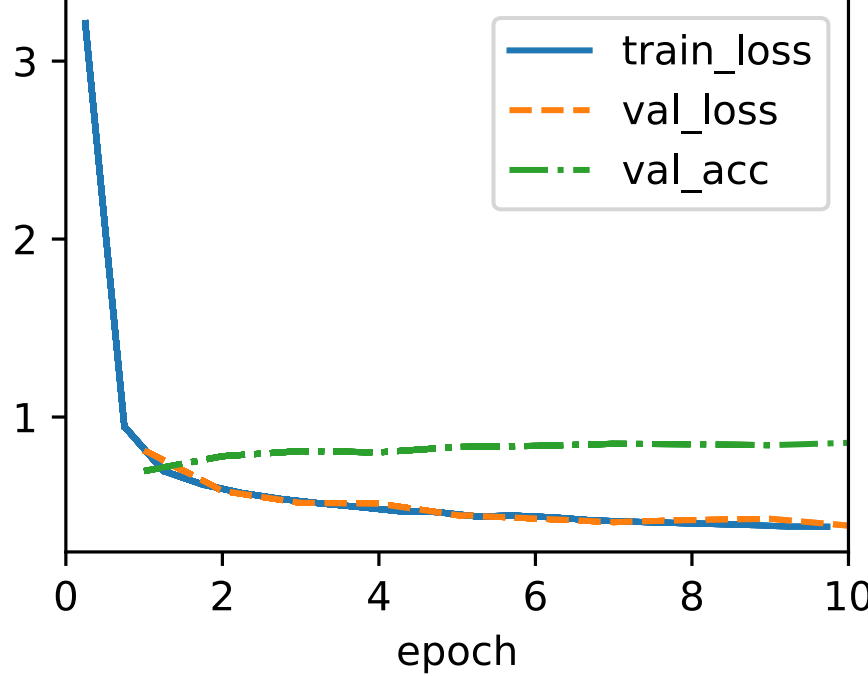

## 11.8.6 Summary and Discussion

You may have noticed that for small datasets like Fashion-MNIST, our implemented vision Transformer does not outperform the ResNet in Section 8.6. Similar observations can be made even on the ImageNet dataset (1.2 million images). This is because Transformers *lack* those useful principles in convolution, such as translation invariance and locality (Section 7.1). However, the picture changes when training larger models on larger datasets (e.g., 300 million images), where vision Transformers outperform ResNets by a large margin in image classification, demonstrating intrinsic superiority of Transformers in scalability (Dosovitskiy *et al.*, 2021). The introduction of vision Transformers has changed the landscape of network design for modeling image data. They were soon shown to be effective on the ImageNet dataset with data-efficient training strategies of DeiT (Touvron *et al.*, 2021). However, the quadratic complexity of self-attention (Section 11.6) makes the Transformer architecture less suitable for higher-resolution images. Towards a general-purpose backbone network in computer vision, Swin Transformers addressed the quadratic computational complexity with respect to image size (Section 11.6.2) and reinstated convolution-like priors, extending the applicability of Transformers to a range of computer vision tasks beyond image classification with state-of-the-art results (Liu *et al.*, 2021).

### 11.8.7 Exercises

1. How does the value of `img_size` affect training time?
2. Instead of projecting the "<cls>" token representation to the output, how would you project the averaged patch representations? Implement this change and see how it affects the accuracy.
3. Can you modify hyperparameters to improve the accuracy of the vision Transformer?

163

Discussions[163].

## 11.9 Large-Scale Pretraining with Transformers

So far in our image classification and machine translation experiments, models have been trained on datasets with input–output examples *from scratch* to perform specific tasks. For example, a Transformer was trained with English–French pairs (Section 11.7) so that this model can translate input English text into French. As a result, each model becomes a *specific expert* that is sensitive to even a slight shift in data distribution (Section 4.7). For better generalized models, or even more competent *generalists* that can perform multiple tasks with or without adaptation, *pretraining* models on large data has been increasingly common.

Given larger data for pretraining, the Transformer architecture performs better with an increased model size and training compute, demonstrating superior *scaling* behavior. Specifically, performance of Transformer-based language models scales as a power law with the amount of model parameters, training tokens, and training compute (Kaplan *et al.*, 2020). The scalability of Transformers is also evidenced by the significantly boosted performance from larger vision Transformers trained on larger data (discussed in Section 11.8). More recent success stories include Gato, a *generalist* model that can play Atari, caption images, chat, and act as a robot (Reed *et al.*, 2022). Gato is a single Transformer that scales well when pretrained on diverse modalities, including text, images, joint torques, and button presses. Notably, all such multimodal data is serialized into a flat sequence of tokens, which can be processed akin to text tokens (Section 11.7) or image patches (Section 11.8) by Transformers.

Prior to the compelling success of pretraining Transformers for multimodal data, Transformers were extensively pretrained with a wealth of text. Originally proposed for machine translation, the Transformer architecture in Fig. 11.7.1 consists of an encoder for representing input sequences and a decoder for generating target sequences. Primarily, Transformers can be used in three different modes: *encoder-only*, *encoder–decoder*, and *decoder-only*. To conclude this chapter, we will review these three modes and explain the scalability in pretraining Transformers.

## 11.9.1 Encoder-Only

When only the Transformer encoder is used, a sequence of input tokens is converted into the same number of representations that can be further projected into output (e.g., classification). A Transformer encoder consists of self-attention layers, where all input tokens attend to each other. For example, vision Transformers depicted in Fig. 11.8.1 are encoder-only, converting a sequence of input image patches into the representation of a special “<cls>” token. Since this representation depends on all input tokens, it is further projected into classification labels. This design was inspired by an earlier encoder-only Transformer pretrained on text: BERT (Bidirectional Encoder Representations from Transformers) (Devlin *et al.*, 2018).

### Pretraining BERT

BERT is pretrained on text sequences using *masked language modeling*: input text with randomly masked tokens is fed into a Transformer encoder to predict the masked tokens. As illustrated in Fig. 11.9.1, an original text sequence “I”, “love”, “this”, “red”, “car” is prepended with the “<cls>” token, and the “<mask>” token randomly replaces “love”; then the cross-entropy loss between the masked token “love” and its prediction is to be minimized during pretraining. Note that there is no constraint in the attention pattern of Transformer encoders (right of Fig. 11.9.1) so all tokens can attend to each other. Thus, prediction of “love” depends on input tokens before and after it in the sequence. This is why BERT is a “bidirectional encoder”. Without need for manual labeling, large-scale text data from books and Wikipedia can be used for pretraining BERT.

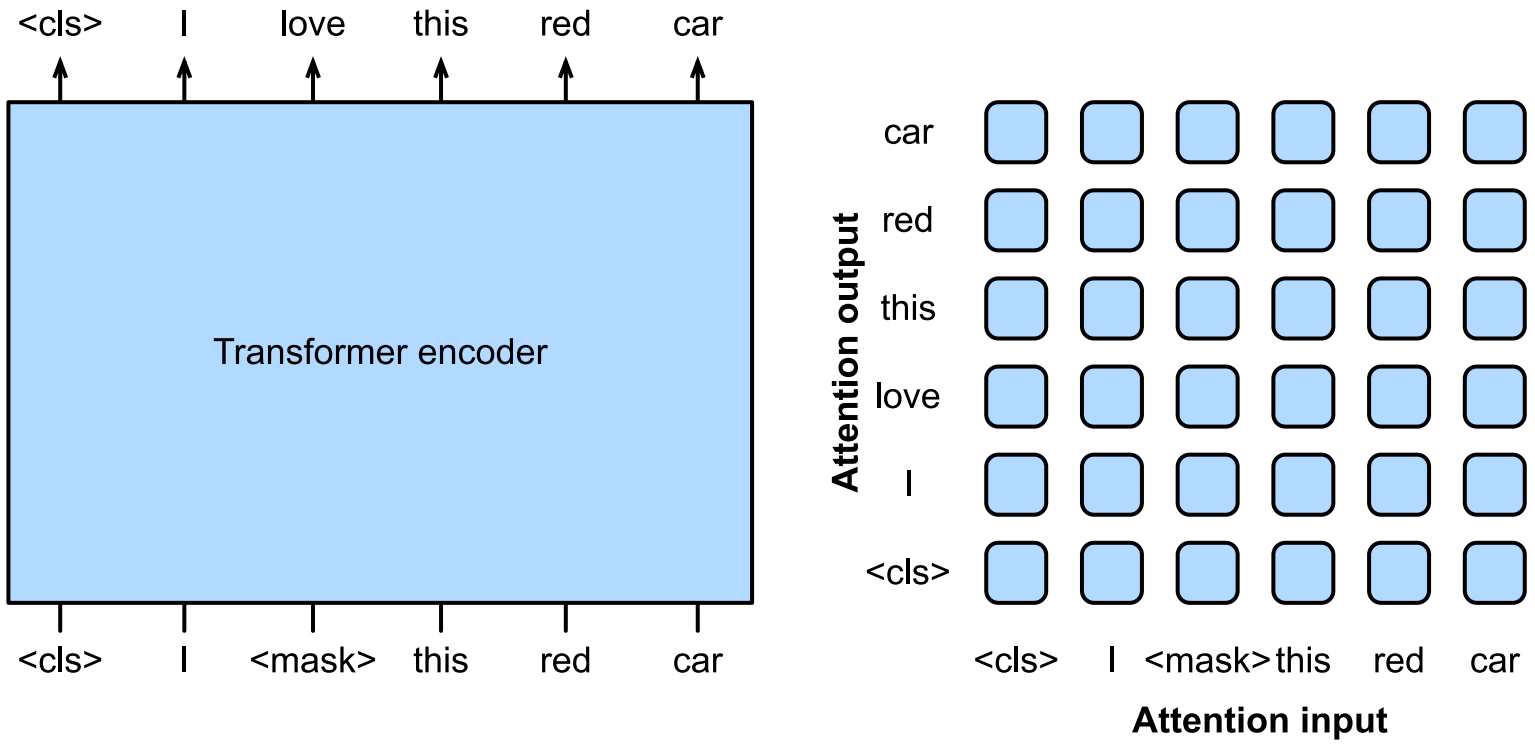

Fig. 11.9.1 Left: Pretraining BERT with masked language modeling. Prediction of the masked "love" token depends on all input tokens before and after "love". Right: Attention pattern in the Transformer encoder. Each token along the vertical axis attends to all input tokens along the horizontal axis.

### Fine-Tuning BERT

The pretrained BERT can be *fine-tuned* to downstream encoding tasks involving single text or text pairs. During fine-tuning, additional layers can be added to BERT with randomized parameters: these parameters and those pretrained BERT parameters will be *updated* to fit training data of downstream tasks.

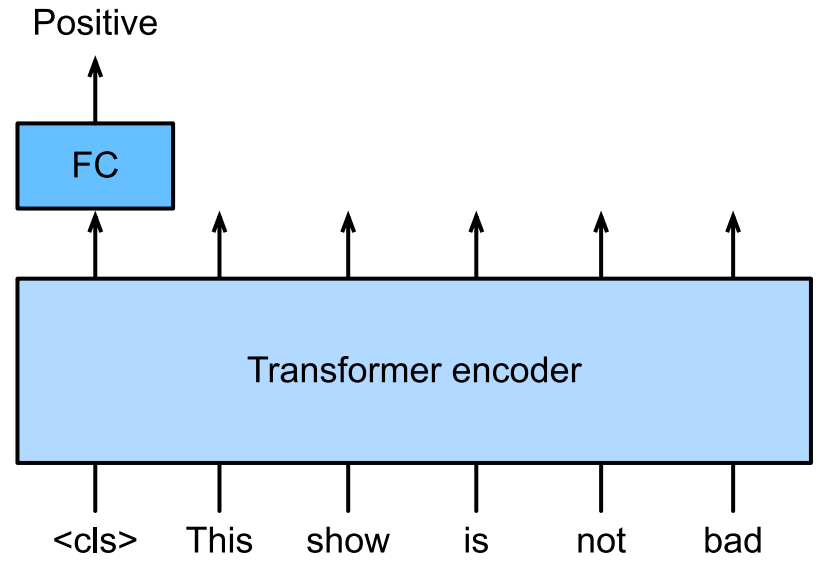

Fig. 11.9.2 Fine-tuning BERT for sentiment analysis.

Fig. 11.9.2 illustrates fine-tuning of BERT for sentiment analysis. The Transformer encoder is a pretrained BERT, which takes a text sequence as input and feeds the "<cls>" representation (global representation of the input) into an additional fully connected layer to predict the sentiment. During fine-tuning, the cross-entropy loss between the prediction and the label on sentiment analysis data is minimized via gradient-based algorithms, where the additional layer is trained from scratch while pretrained parameters of BERT are updated. BERT does more than sentiment analysis. The general language representations learned by the 350-million-parameter BERT from 250 billion training tokens advanced the state of the art for natural language tasks such as single text classification, text pair classification or regression, text tagging, and question answering.

You may note that these downstream tasks include text pair understanding. BERT pretraining has another loss for predicting whether one sentence immediately follows the other. However, this loss was later found to be less useful when pretraining RoBERTa, a BERT variant of the same size, on 2000 billion tokens (Liu *et al.*, 2019). Other derivatives of BERT improved model architectures or pretraining objectives, such as ALBERT (enforcing parameter sharing) (Lan *et al.*, 2019), SpanBERT (representing and predicting spans of text) (Joshi *et al.*, 2020), DistilBERT (lightweight via knowledge distillation) (Sanh *et al.*, 2019), and ELECTRA (replaced token detection) (Clark *et al.*, 2020). Moreover, BERT inspired Transformer pretraining in computer vision, such as with vision Transformers (Dosovitskiy *et al.*, 2021), Swin Transformers (Liu *et al.*, 2021), and MAE (masked autoencoders) (He *et al.*, 2022).

## 11.9.2 Encoder–Decoder

Since a Transformer encoder converts a sequence of input tokens into the same number of output representations, the encoder-only mode cannot generate a sequence of arbitrary length as in machine translation. As originally proposed for machine translation, the Transformer architecture can be outfitted with a decoder that autoregressively predicts the target sequence of arbitrary length, token by token, conditional on both encoder output and decoder output: (i) for conditioning on encoder output, encoder–decoder cross-attention (multi-head attention of decoder in Fig. 11.7.1) allows target tokens to attend to *all* input tokens; (ii) conditioning on decoder output is achieved by a so-called *causal* attention (this name is common in the literature but is misleading as it has little connection to the proper study of causality) pattern (masked multi-head attention of decoder in Fig. 11.7.1), where any target token can only attend to *past* and *present* tokens in the target sequence.

To pretrain encoder–decoder Transformers beyond human-labeled machine translation data, BART (Lewis *et al.*, 2019) and T5 (Raffel *et al.*, 2020) are two concurrently proposed encoder–decoder Transformers pretrained on large-scale text corpora. Both attempt to reconstruct original text in their pretraining objectives, while the former emphasizes noising input (e.g., masking, deletion, permutation, and rotation) and the latter highlights multitask unification with comprehensive ablation studies.

### Pretraining T5

As an example of the pretrained Transformer encoder–decoder, T5 (Text-to-Text Transfer Transformer) unifies many tasks as the same text-to-text problem: for any task, the input of the encoder is a task description (e.g., "Summarize", ":") followed by task input (e.g., a sequence of tokens from an article), and the decoder predicts the task output (e.g., a sequence of tokens summarizing the input article). To perform as text-to-text, T5 is trained to generate some target text conditional on input text.

To obtain input and output from any original text, T5 is pretrained to predict consecutive spans. Specifically, tokens from text are randomly replaced by special tokens where each consecutive span is replaced by the same special token. Consider the example in Fig. 11.9.3, where the original text is “I”, “love”, “this”, “red”, “car”. Tokens “love”, “red”, “car” are randomly replaced by special tokens. Since “red” and “car” are a consecutive span, they are replaced by the same special token. As a result, the input sequence is “I”, “<X>”, “this”, “<Y>”, and the target sequence is “<X>”, “love”, “<Y>”, “red”, “car”, “<Z>”, where “<Z>” is another special token marking the end. As shown in Fig. 11.9.3, the decoder has a causal attention pattern to prevent itself from attending to future tokens during sequence prediction.

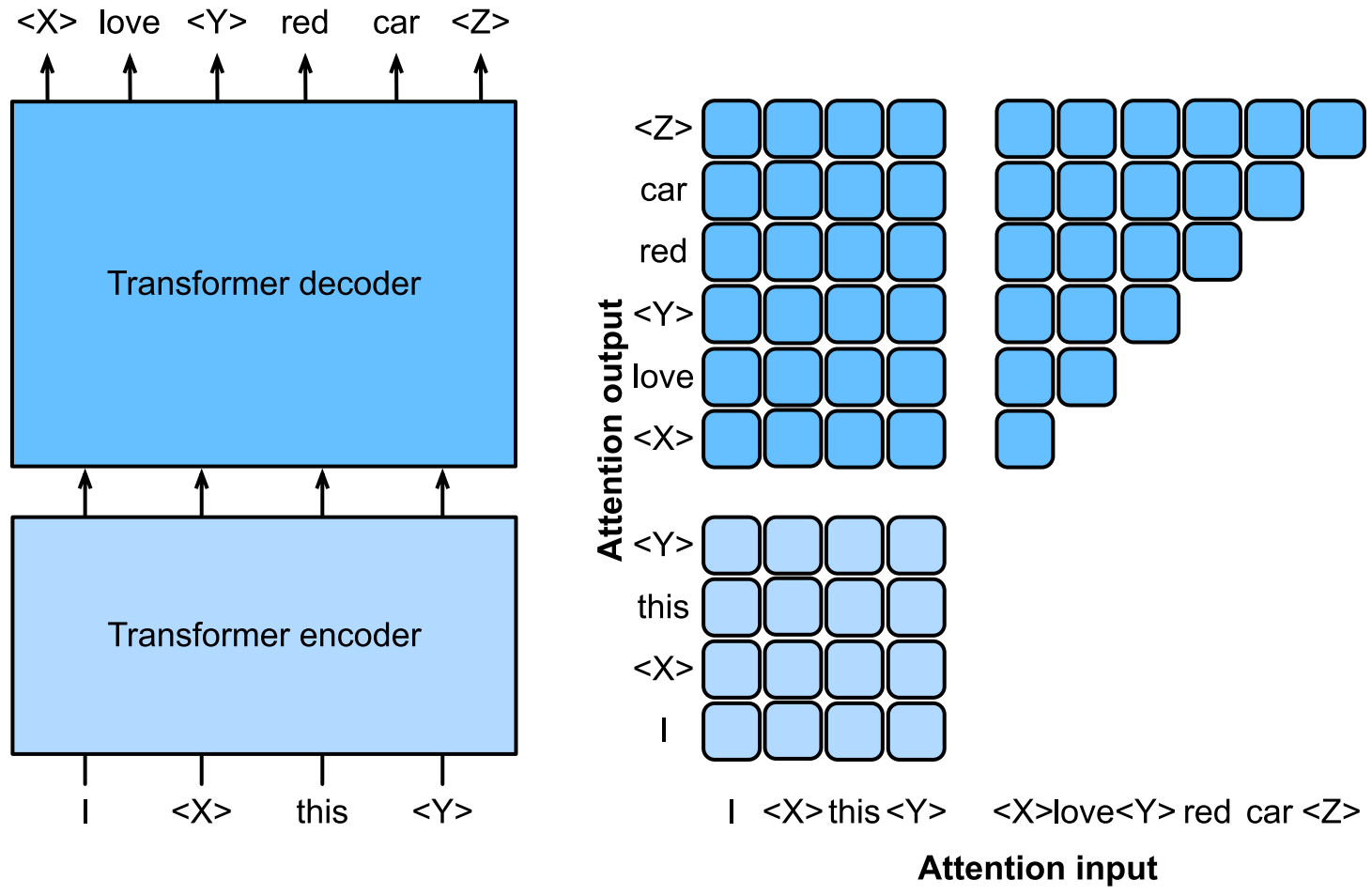

Fig. 11.9.3 Left: Pretraining T5 by predicting consecutive spans. The original sentence is “I”, “love”, “this”, “red”, “car”, where “love” is replaced by a special “<X>” token, and consecutive “red”, “car” are replaced by a special “<Y>” token. The target sequence ends with a special “<Z>” token. Right: Attention pattern in the Transformer encoder–decoder. In the encoder self-attention (lower square), all input tokens attend to each other; In the encoder–decoder cross-attention (upper rectangle), each target token attends to all input tokens; In the decoder self-attention (upper triangle), each target token attends to present and past target tokens only (causal).

In T5, predicting consecutive span is also referred to as reconstructing corrupted text. With this objective, T5 is pretrained with 1000 billion tokens from the C4 (Colossal Clean Crawled Corpus) data, which consists of clean English text from the web (Raffel *et al.*, 2020).

### Fine-Tuning T5

Similar to BERT, T5 needs to be fine-tuned (updating T5 parameters) on task-specific training data to perform this task. Major differences from BERT fine-tuning include: (i) T5 input includes task descriptions; (ii) T5 can generate sequences with arbitrary length with its Transformer decoder; (iii) No additional layers are required.

Fig. 11.9.4 explains fine-tuning T5 using text summarization as an example. In this down-stream task, the task description tokens “Summarize”, “:” followed by the article tokens are input to the encoder.

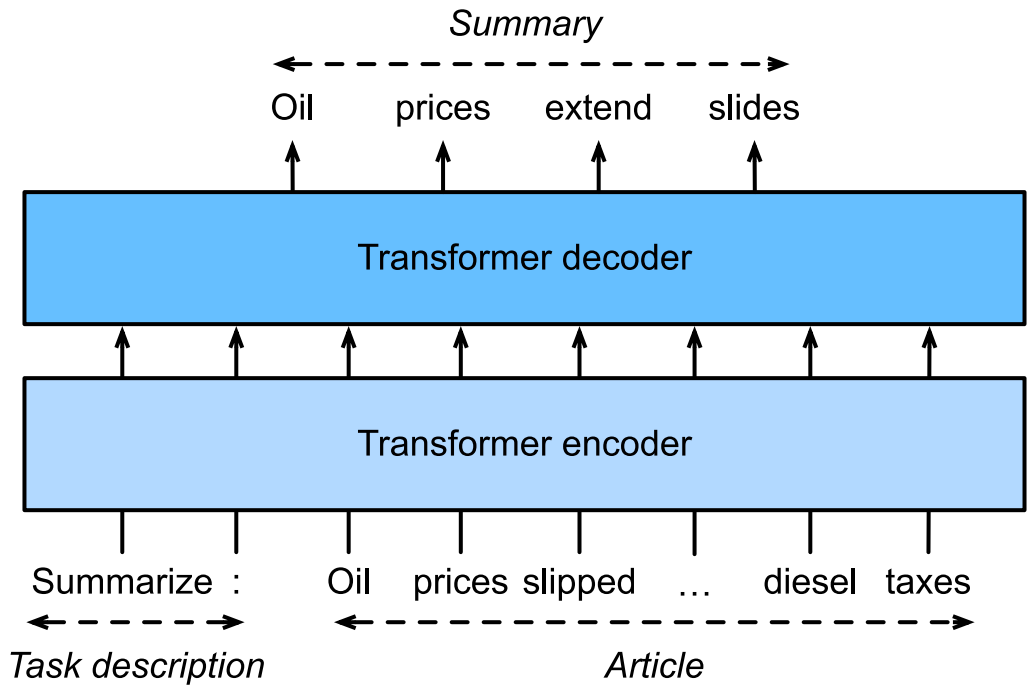

Fig. 11.9.4 Fine-tuning T5 for text summarization. Both the task description and article tokens are fed into the Transformer encoder for predicting the summary.

After fine-tuning, the 11-billion-parameter T5 (T5-11B) achieved state-of-the-art results on multiple encoding (e.g., classification) and generation (e.g., summarization) benchmarks. Since released, T5 has been extensively used in later research. For example, switch Transformers are designed based on T5 to activate a subset of the parameters for better computational efficiency (Fedus *et al.*, 2022). In a text-to-image model called Imagen, text is input to a frozen T5 encoder (T5-XXL) with 4.6 billion parameters (Saharia *et al.*, 2022). The photorealistic text-to-image examples in Fig. 11.9.5 suggest that the T5 encoder alone may effectively represent text even without fine-tuning.

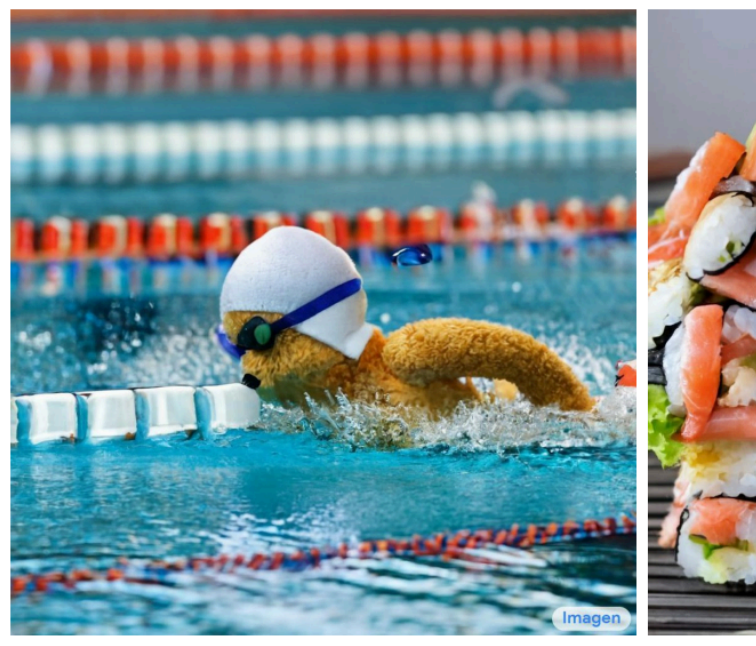

Teddy bears swimming at the Olympics 400m Butterfly event.

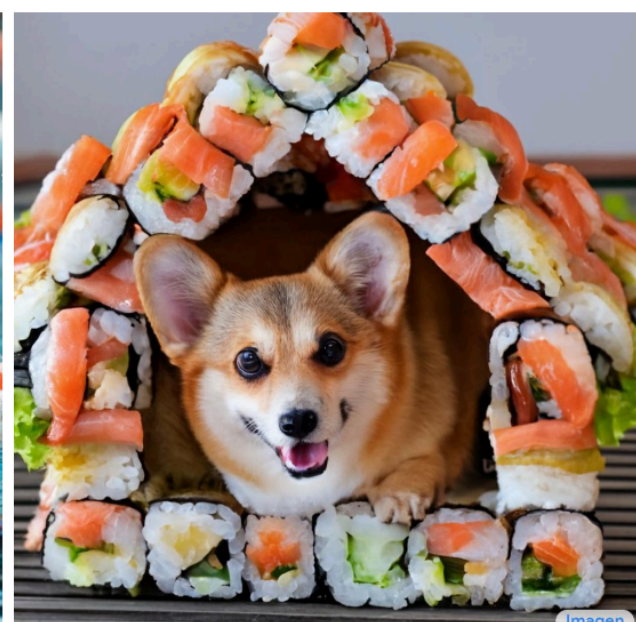

A cute corgi lives in a house made out of sushi.

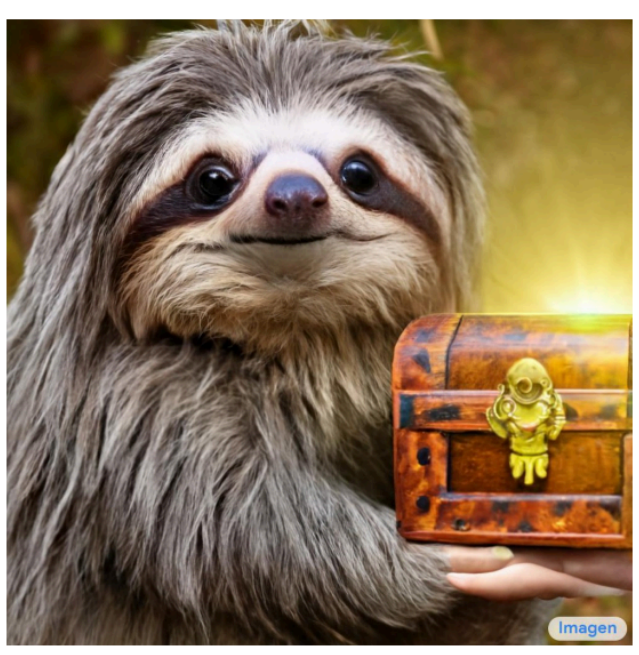

A cute sloth holding a small treasure chest. A bright golden glow is coming from the chest.

Fig. 11.9.5 Text-to-image examples by the Imagen model, whose text encoder is from T5 (figures taken from Saharia et al. (2022)).

### 11.9.3 Decoder-Only

We have reviewed encoder-only and encoder–decoder Transformers. Alternatively, decoder-only Transformers remove the entire encoder and the decoder sublayer with the encoder–decoder cross-attention from the original encoder–decoder architecture depicted in Fig. 11.7.1.

Nowadays, decoder-only Transformers have been the *de facto* architecture in large-scale language modeling (Section 9.3), which leverages the world's abundant unlabeled text corpora via self-supervised learning.

## GPT and GPT-2

Using language modeling as the training objective, the GPT (generative pre-training) model chooses a Transformer decoder as its backbone (Radford *et al.*, 2018).

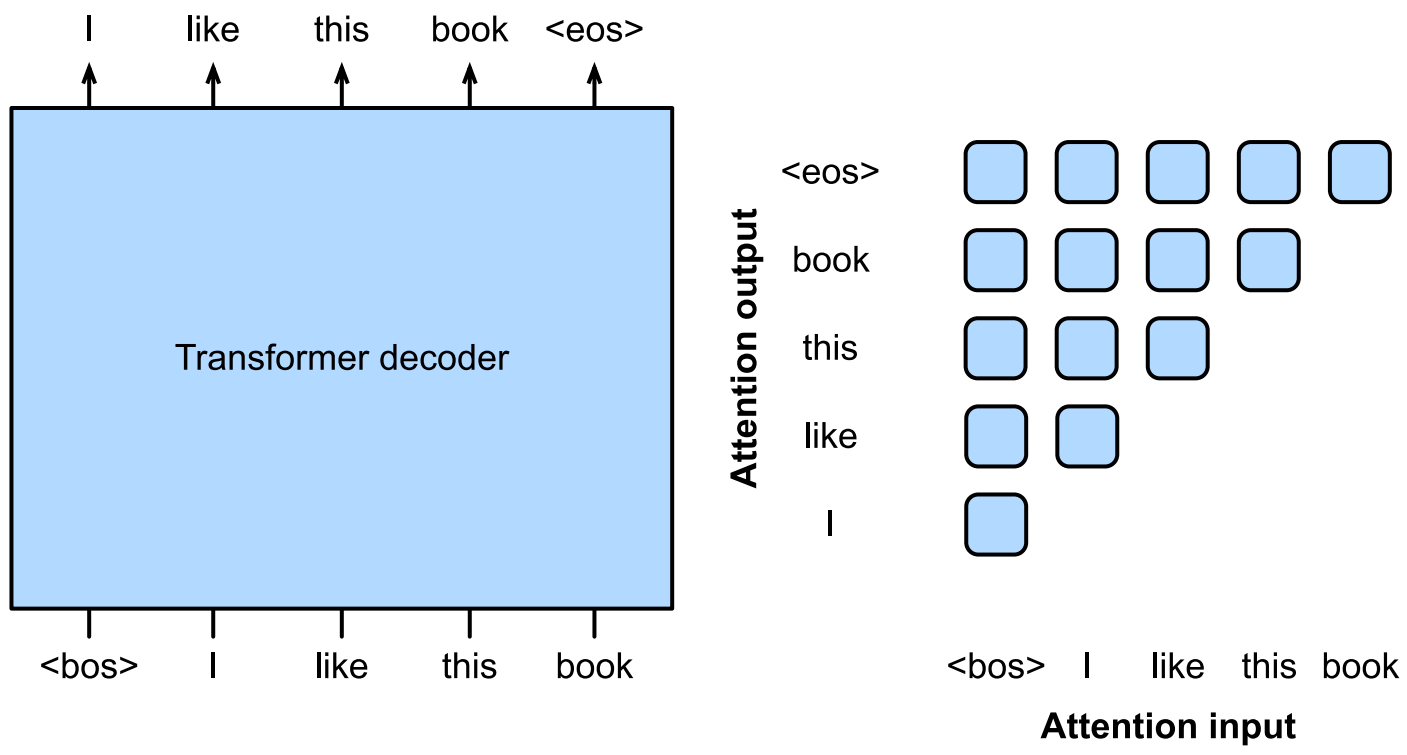

Fig. 11.9.6 Left: Pretraining GPT with language modeling. The target sequence is the input sequence shifted by one token. Both "<bos>" and "<eos>" are special tokens marking the beginning and end of sequences, respectively. Right: Attention pattern in the Transformer decoder. Each token along the vertical axis attends to only its past tokens along the horizontal axis (causal).

Following the autoregressive language model training as described in Section 9.3.3, Fig. 11.9.6 illustrates GPT pretraining with a Transformer encoder, where the target sequence is the input sequence shifted by one token. Note that the attention pattern in the Transformer decoder enforces that each token can only attend to its past tokens (future tokens cannot be attended to because they have not yet been chosen).

GPT has 100 million parameters and needs to be fine-tuned for individual downstream tasks. A much larger Transformer-decoder language model, GPT-2, was introduced one year later (Radford *et al.*, 2019). Compared with the original Transformer decoder in GPT, pre-normalization (discussed in Section 11.8.3) and improved initialization and weight-scaling were adopted in GPT-2. Pretrained on 40 GB of text, the 1.5-billion-parameter GPT-2 obtained the state-of-the-art results on language modeling benchmarks and promising results on multiple other tasks *without updating the parameters or architecture*.

## GPT-3 and Beyond

GPT-2 demonstrated potential of using the same language model for multiple tasks without updating the model. This is more computationally efficient than fine-tuning, which requires model updates via gradient computation.

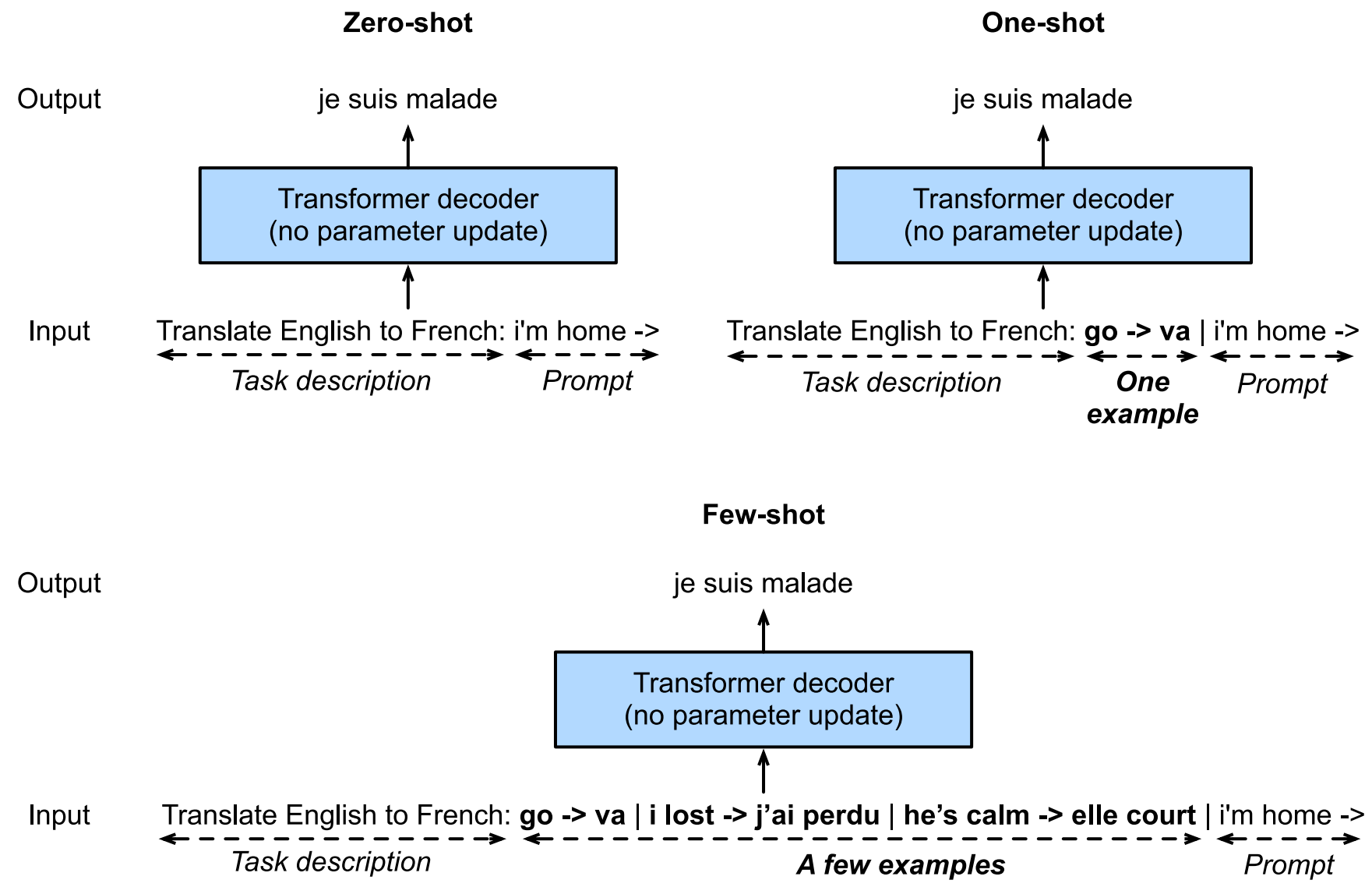

Fig. 11.9.7 Zero-shot, one-shot, few-shot in-context learning with language models (Transformer decoders). No parameter update is needed.

Before explaining the more computationally efficient use of language models without parameter update, recall Section 9.5 that a language model can be trained to generate a text sequence conditional on some prefix text sequence. Thus, a pretrained language model may generate the task output as a sequence *without parameter update*, conditional on an input sequence with the task description, task-specific input–output examples, and a prompt (task input). This learning paradigm is called *in-context learning* (Brown *et al.*, 2020), which can be further categorized into *zero-shot*, *one-shot*, and *few-shot*, when there is no, one, and a few task-specific input–output examples (Fig. 11.9.7).

These three settings were tested in GPT-3 (Brown *et al.*, 2020), whose largest version uses data and model size about two orders of magnitude larger than those in GPT-2. GPT-3 uses the same Transformer decoder architecture as its direct predecessor GPT-2 except that attention patterns (at the right in Fig. 11.9.6) are sparser at alternating layers. Pretrained with 300 billion tokens, GPT-3 performs better with larger model size, where few-shot performance increases most rapidly (Fig. 11.9.8).

The subsequent GPT-4 model did not fully disclose technical details in its report (OpenAI, 2023). By contrast with its predecessors, GPT-4 is a large-scale, multimodal model that can take both text and images as input and generate text output.

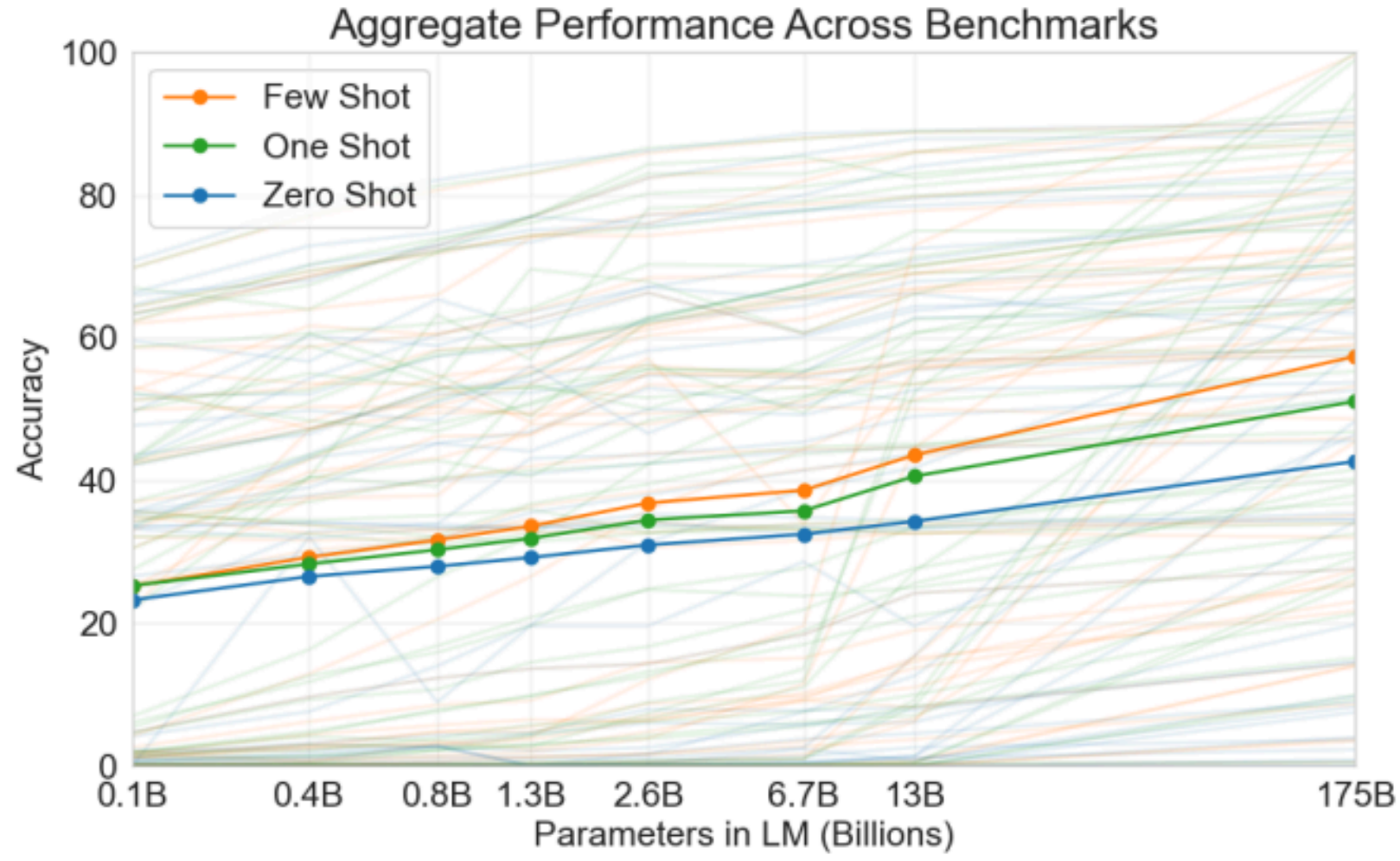

Fig. 11.9.8 Aggregate performance of GPT-3 for all 42 accuracy-denominated benchmarks (caption adapted and figure taken from Brown et al. (2020)).

## 11.9.4 Scalability

Fig. 11.9.8 empirically demonstrates scalability of Transformers in the GPT-3 language model. For language modeling, more comprehensive empirical studies on the scalability of Transformers have led researchers to see promise in training larger Transformers with more data and compute (Kaplan *et al.*, 2020).

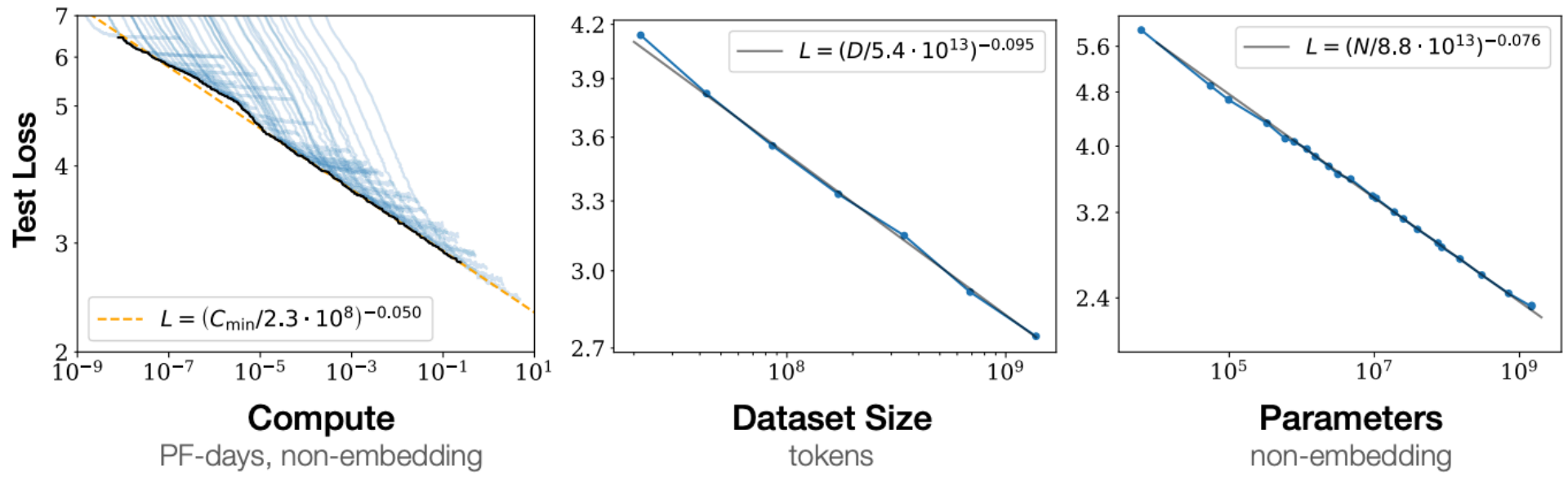

Fig. 11.9.9 Transformer language model performance improves smoothly as we increase the model size, dataset size, and amount of compute used for training. For optimal performance all three factors must be scaled up in tandem. Empirical performance has a power-law relationship with each individual factor when not bottlenecked by the other two (caption adapted and figure taken from Kaplan et al. (2020)).

As shown in Fig. 11.9.9, *power-law scaling* can be observed in the performance with respect to the model size (number of parameters, excluding embedding layers), dataset size (number of training tokens), and amount of training compute (PetaFLOP/s-days, excluding embedding layers). In general, increasing all these three factors in tandem leads to better performance. However, *how* to increase them in tandem still remains a matter of debate (Hoffmann *et al.*, 2022).

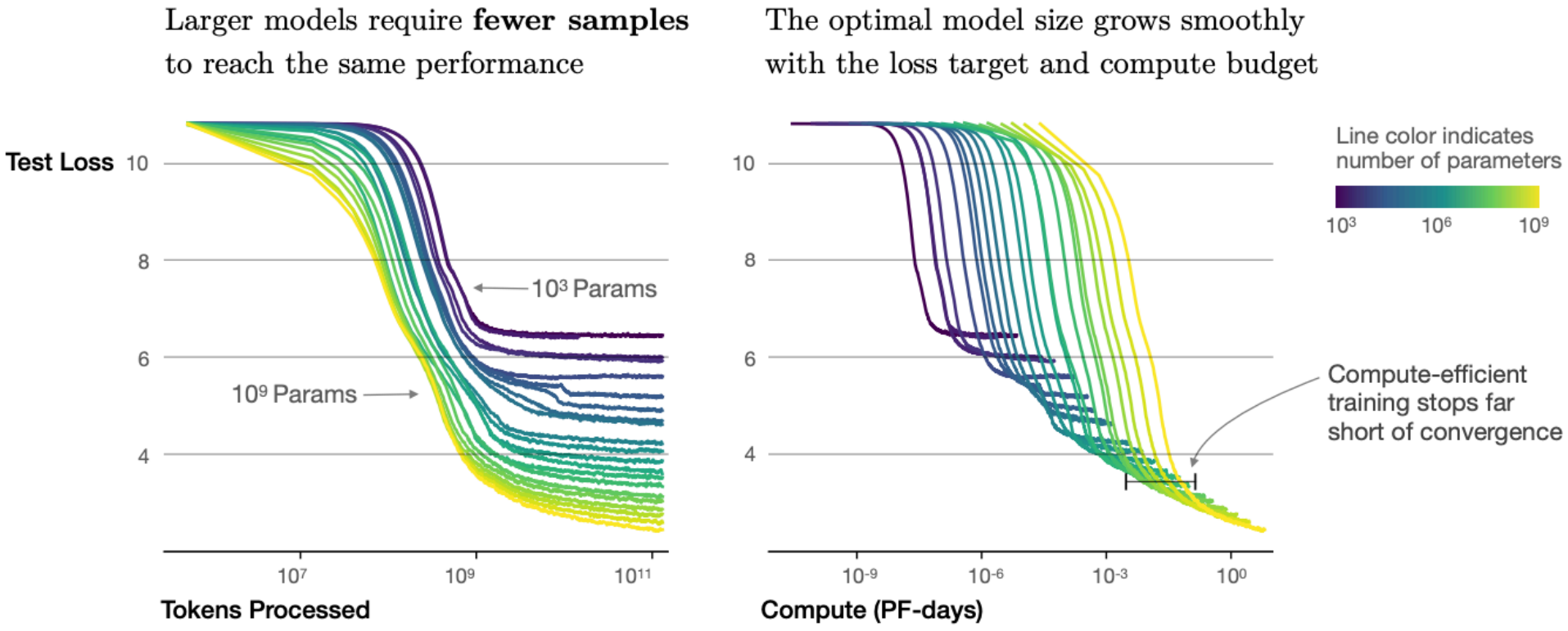

Fig. 11.9.10 Transformer language model training runs (figure taken from Kaplan et al. (2020)).

As well as increased performance, large models also enjoy better sample efficiency than small models. Fig. 11.9.10 shows that large models need fewer training samples (tokens processed) to perform at the same level achieved by small models, and performance is scaled smoothly with compute.

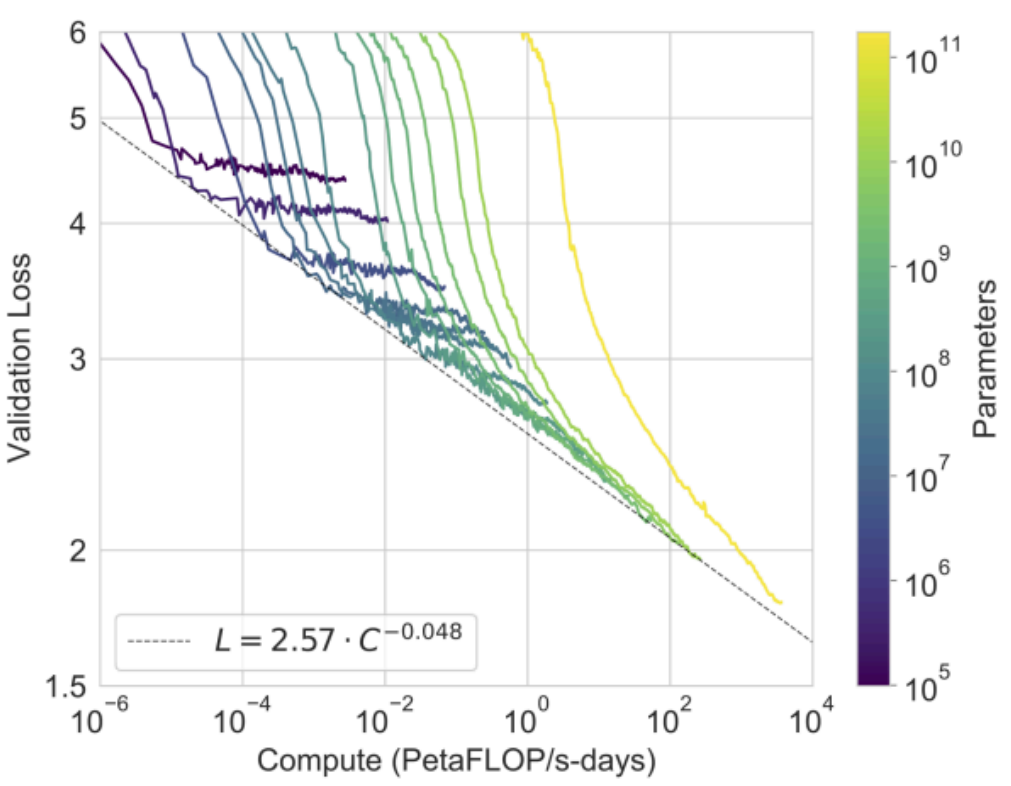

Fig. 11.9.11 GPT-3 performance (cross-entropy validation loss) follows a power-law trend with the amount of compute used for training. The power-law behavior observed in Kaplan et al. (2020) continues for an additional two orders of magnitude with only small deviations from the predicted curve. Embedding parameters are excluded from compute and parameter counts (caption adapted and figure taken from Brown et al. (2020)).

The empirical scaling behaviors in Kaplan *et al.* (2020) have been tested in subsequent large Transformer models. For example, GPT-3 supported this hypothesis with two more orders of magnitude in Fig. 11.9.11.

## 11.9.5 Large Language Models

The scalability of Transformers in the GPT series has inspired subsequent large language models. The GPT-2 Transformer decoder was used for training the 530-billion-parameter Megatron-Turing NLG (Smith *et al.*, 2022) with 270 billion training tokens. Following the GPT-2 design, the 280-billion-parameter Gopher (Rae *et al.*, 2021) pretrained with 300 billion tokens, performed competitively across diverse tasks. Inheriting the same architecture and using the same compute budget of Gopher, Chinchilla (Hoffmann *et al.*, 2022) is a substantially smaller (70 billion parameters) model that trains for much longer (1.4 trillion training tokens), outperforming Gopher on many tasks and with more emphasis on the number of tokens than on the number of parameters. To continue the scaling line of language modeling, PaLM (Pathway Language Model) (Chowdhery *et al.*, 2022), a 540-billion-parameter Transformer decoder with modified designs pretrained on 780 billion tokens, outperformed average human performance on the BIG-Bench benchmark (Srivastava *et al.*, 2022). Its later version, PaLM 2 (Anil *et al.*, 2023), scaled data and model roughly 1:1 and improved multilingual and reasoning capabilities. Other large language models, such as Minerva (Lewkowycz *et al.*, 2022) that further trains a generalist (PaLM) and Galactica (Taylor *et al.*, 2022) that is not trained on a general corpus, have shown promising quantitative and scientific reasoning capabilities.

Open-sourced releases, such as OPT (Open Pretrained Transformers) (Zhang *et al.*, 2022), BLOOM (Scao *et al.*, 2022), and FALCON (Penedo *et al.*, 2023), democratized research and use of large language models. Focusing on computational efficiency at inference time, the open-sourced Llama 1 (Touvron *et al.*, 2023a) outperformed much larger models by training on more tokens than had been typically used. The updated Llama 2 (Touvron *et al.*, 2023b) further increased the pretraining corpus by 40%, leading to product models that may match the performance of competitive close-sourced models.

Wei *et al.* (2022) discussed emergent abilities of large language models that are present in larger models, but not in smaller models. However, simply increasing model size does not inherently make models follow human instructions better. Sanh *et al.* (2021), Wei *et al.* (2021) have found that fine-tuning large language models on a range of datasets described via *instructions* can improve zero-shot performance on held-out tasks. Using *reinforcement learning from human feedback*, Ouyang *et al.* (2022) fine-tuned GPT-3 to follow a diverse set of instructions. Following the resultant InstructGPT which aligns language models with human intent via fine-tuning (Ouyang *et al.*, 2022), ChatGPT[164] can generate human-like responses (e.g., code debugging and creative writing) based on conversations with humans and can perform many natural language processing tasks zero-shot (Qin *et al.*, 2023). Bai *et al.* (2022) replaced human inputs (e.g., human-labeled data) with model outputs to partially automate the instruction tuning process, which is also known as *reinforcement learning from AI feedback*.

Large language models offer an exciting prospect of formulating text input to induce models to perform desired tasks via in-context learning, which is also known as *prompting*. Notably, *chain-of-thought prompting* (Wei *et al.*, 2022), an in-context learning method with few-shot "question, intermediate reasoning steps, answer" demonstrations, elicits the complex reasoning capabilities of large language models in order to solve mathematical, commonsense, and symbolic reasoning tasks. Sampling multiple reasoning paths (Wang *et al.*, 2023), diversifying few-shot demonstrations (Zhang *et al.*, 2023), and reducing complex problems to sub-problems (Zhou *et al.*, 2023) can all improve the reasoning accuracy. In fact, with simple prompts like "Let's think step by step" just before each answer, large language models can even perform *zero-shot* chain-of-thought reasoning with decent accuracy (Kojima *et al.*, 2022). Even for multimodal inputs consisting of both text and images, language models can perform multimodal chain-of-thought reasoning with higher accuracy than using text input only (Zhang *et al.*, 2023).

## 11.9.6 Summary and Discussion

Transformers have been pretrained as encoder-only (e.g., BERT), encoder–decoder (e.g., T5), and decoder-only (e.g., GPT series). Pretrained models may be adapted to perform different tasks with model update (e.g., fine-tuning) or not (e.g., few-shot). Scalability of Transformers suggests that better performance benefits from larger models, more training data, and more training compute. Since Transformers were first designed and pretrained for text data, this section leans slightly towards natural language processing. Nonetheless, those models discussed above can be often found in more recent models across multiple modalities. For example, (i) Chinchilla (Hoffmann *et al.*, 2022) was further extended to Flamingo (Alayrac *et al.*, 2022), a visual language model for few-shot learning; (ii) GPT-2 (Radford *et al.*, 2019) and the vision Transformer encode text and images in CLIP (Contrastive Language-Image Pre-training) (Radford *et al.*, 2021), whose image and text embeddings were later adopted in the DALL-E 2 text-to-image system (Ramesh *et al.*, 2022). Although there have been no systematic studies on Transformer scalability in multimodal pretraining yet, an all-Transformer text-to-image model called Parti (Yu *et al.*, 2022) shows potential of scalability across modalities: a larger Parti is more capable of high-fidelity image generation and content-rich text understanding (Fig. 11.9.12).

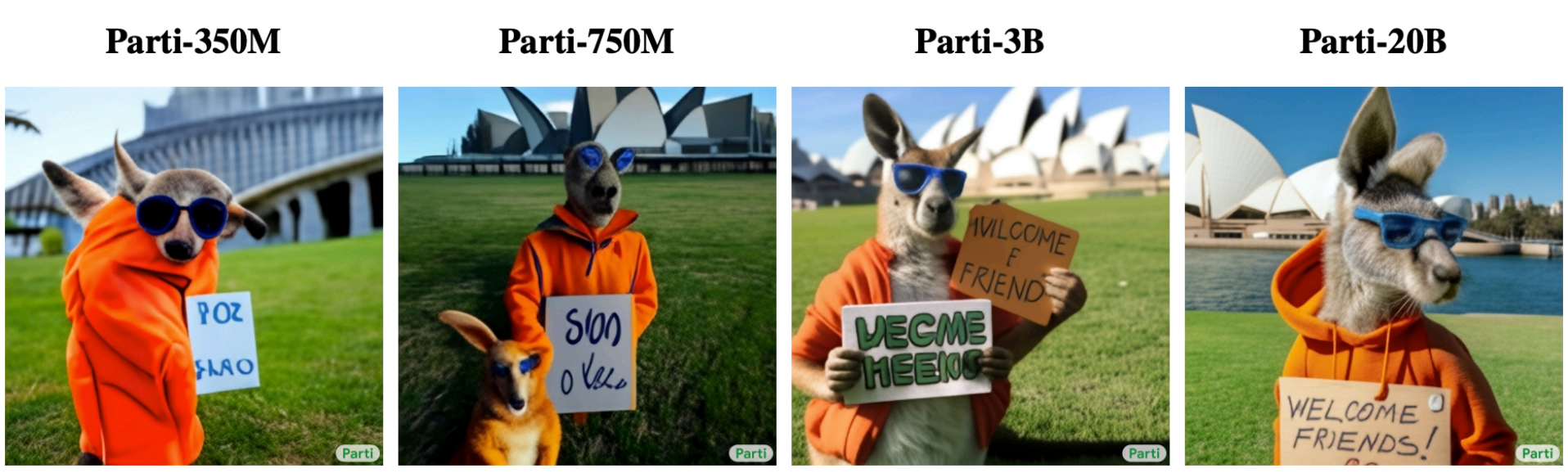

A portrait photo of a kangaroo wearing an orange hoodie and blue sunglasses standing on the grass in front of the Sydney Opera House holding a sign on the chest that says Welcome Friends!

Fig. 11.9.12 Image examples generated from the same text by the Parti model of increasing sizes (350M, 750M, 3B, 20B) (examples taken from Yu et al. (2022)).

## 11.9.7 Exercises

1. Is it possible to fine-tune T5 using a minibatch consisting of different tasks? Why or why not? How about for GPT-2?
2. Given a powerful language model, what applications can you think of?
3. Say that you are asked to fine-tune a language model to perform text classification by adding additional layers. Where will you add them? Why?
4. Consider sequence-to-sequence problems (e.g., machine translation) where the input sequence is always available throughout the target sequence prediction. What could be limitations of modeling with decoder-only Transformers? Why?

Discussions[165].

# 12 Optimization Algorithms

If you read the book in sequence up to this point you already used a number of optimization algorithms to train deep learning models. They were the tools that allowed us to continue updating model parameters and to minimize the value of the loss function, as evaluated on the training set. Indeed, anyone content with treating optimization as a black box device to minimize objective functions in a simple setting might well content oneself with the knowledge that there exists an array of incantations of such a procedure (with names such as "SGD" and "Adam").

To do well, however, some deeper knowledge is required. Optimization algorithms are important for deep learning. On the one hand, training a complex deep learning model can take hours, days, or even weeks. The performance of the optimization algorithm directly affects the model's training efficiency. On the other hand, understanding the principles of different optimization algorithms and the role of their hyperparameters will enable us to tune the hyperparameters in a targeted manner to improve the performance of deep learning models.

In this chapter, we explore common deep learning optimization algorithms in depth. Almost all optimization problems arising in deep learning are *nonconvex*. Nonetheless, the design and analysis of algorithms in the context of *convex* problems have proven to be very instructive. It is for that reason that this chapter includes a primer on convex optimization and the proof for a very simple stochastic gradient descent algorithm on a convex objective function.

## 12.1 Optimization and Deep Learning

In this section, we will discuss the relationship between optimization and deep learning as well as the challenges of using optimization in deep learning. For a deep learning problem, we will usually define a *loss function* first. Once we have the loss function, we can use an optimization algorithm in attempt to minimize the loss. In optimization, a loss function is often referred to as the *objective function* of the optimization problem. By tradition and convention most optimization algorithms are concerned with *minimization*. If we ever need to maximize an objective there is a simple solution: just flip the sign on the objective.

### 12.1.1 Goal of Optimization

Although optimization provides a way to minimize the loss function for deep learning, in essence, the goals of optimization and deep learning are fundamentally different. The former is primarily concerned with minimizing an objective whereas the latter is concerned with finding a suitable model, given a finite amount of data. In Section 3.6, we discussed the difference between these two goals in detail. For instance, training error and generalization error generally differ: since the objective function of the optimization algorithm is usually a loss function based on the training dataset, the goal of optimization is to reduce the training error. However, the goal of deep learning (or more broadly, statistical inference) is to reduce the generalization error. To accomplish the latter we need to pay attention to overfitting in addition to using the optimization algorithm to reduce the training error.

```
%matplotlib inline
import numpy as np
import torch
from mpl_toolkits import mplot3d
from d2l import torch as d2l
```

To illustrate the aforementioned different goals, let's consider the empirical risk and the risk. As described in Section 4.7.3, the empirical risk is an average loss on the training dataset while the risk is the expected loss on the entire population of data. Below we define two functions: the risk function `f` and the empirical risk function `g`. Suppose that we have only a finite amount of training data. As a result, here `g` is less smooth than `f`.

```
def f(x):
    return x * torch.cos(np.pi * x)

def g(x):
    return f(x) + 0.2 * torch.cos(5 * np.pi * x)
```

The graph below illustrates that the minimum of the empirical risk on a training dataset may be at a different location from the minimum of the risk (generalization error).

```
def annotate(text, xy, xytext):  #@save
    d2l.plt.gca().annotate(text, xy=xy, xytext=xytext,
                           arrowprops=dict(arrowstyle='->'))

x = torch.arange(0.5, 1.5, 0.01)
d2l.set_figsize((4.5, 2.5))
d2l.plot(x, [f(x), g(x)], 'x', 'risk')
annotate('min of\nempirical risk', (1.0, -1.2), (0.5, -1.1))
annotate('min of risk', (1.1, -1.05), (0.95, -0.5))
```

### 12.1.2 Optimization Challenges in Deep Learning

In this chapter, we are going to focus specifically on the performance of optimization algorithms in minimizing the objective function, rather than a model's generalization error. In

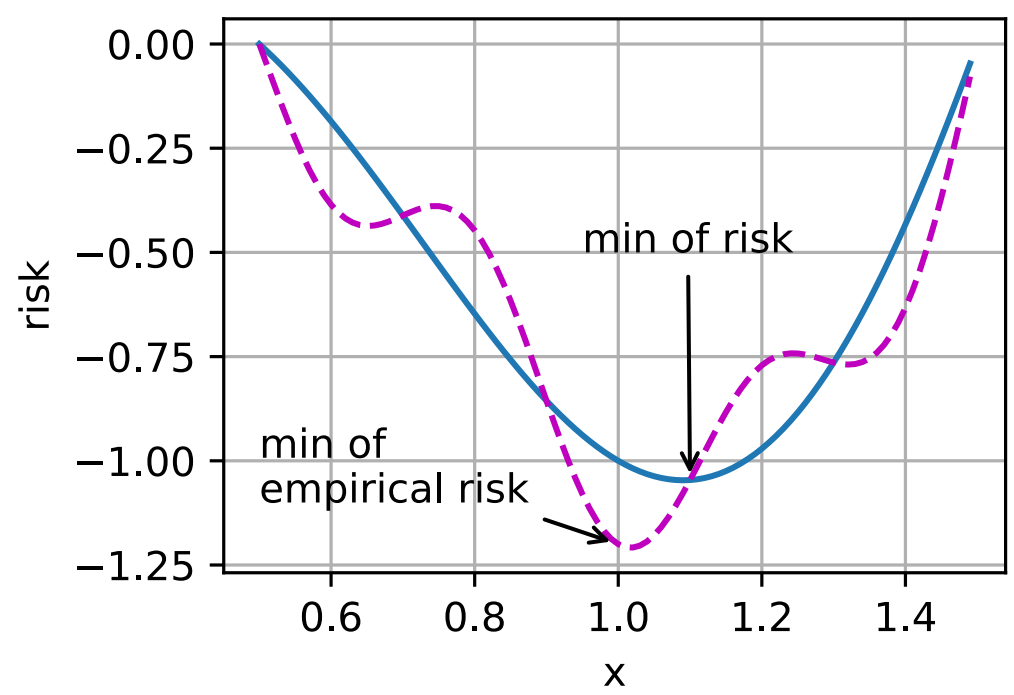

Section 3.1 we distinguished between analytical solutions and numerical solutions in optimization problems. In deep learning, most objective functions are complicated and do not have analytical solutions. Instead, we must use numerical optimization algorithms. The optimization algorithms in this chapter all fall into this category.

There are many challenges in deep learning optimization. Some of the most vexing ones are local minima, saddle points, and vanishing gradients. Let's have a look at them.

### Local Minima

For any objective function $f(x)$, if the value of $f(x)$ at $x$ is smaller than the values of $f(x)$ at any other points in the vicinity of $x$, then $f(x)$ could be a local minimum. If the value of $f(x)$ at $x$ is the minimum of the objective function over the entire domain, then $f(x)$ is the global minimum.

For example, given the function

$$f(x) = x \cdot \cos(\pi x) \text{ for } -1.0 \leq x \leq 2.0, \tag{12.1.1}$$

we can approximate the local minimum and global minimum of this function.

```
x = torch.arange(-1.0, 2.0, 0.01)
d2l.plot(x, [f(x), ], 'x', 'f(x)')
annotate('local minimum', (-0.3, -0.25), (-0.77, -1.0))
annotate('global minimum', (1.1, -0.95), (0.6, 0.8))
```

The objective function of deep learning models usually has many local optima. When the numerical solution of an optimization problem is near the local optimum, the numerical solution obtained by the final iteration may only minimize the objective function *locally*, rather than *globally*, as the gradient of the objective function's solutions approaches or becomes zero. Only some degree of noise might knock the parameter out of the local minimum. In fact, this is one of the beneficial properties of minibatch stochastic gradient descent where the natural variation of gradients over minibatches is able to dislodge the parameters from local minima.

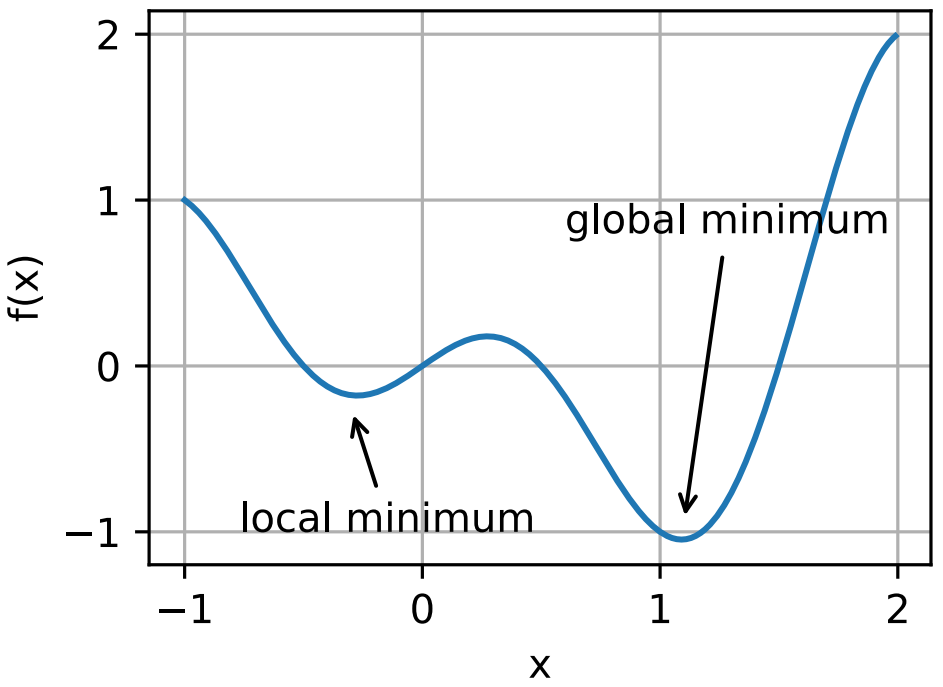

### Saddle Points

Besides local minima, saddle points are another reason for gradients to vanish. A *saddle point* is any location where all gradients of a function vanish but which is neither a global nor a local minimum. Consider the function $f(x) = x^3$. Its first and second derivative vanish for $x = 0$. Optimization might stall at this point, even though it is not a minimum.

```
x = torch.arange(-2.0, 2.0, 0.01)
d2l.plot(x, [x**3], 'x', 'f(x)')
annotate('saddle point', (0, -0.2), (-0.52, -5.0))
```

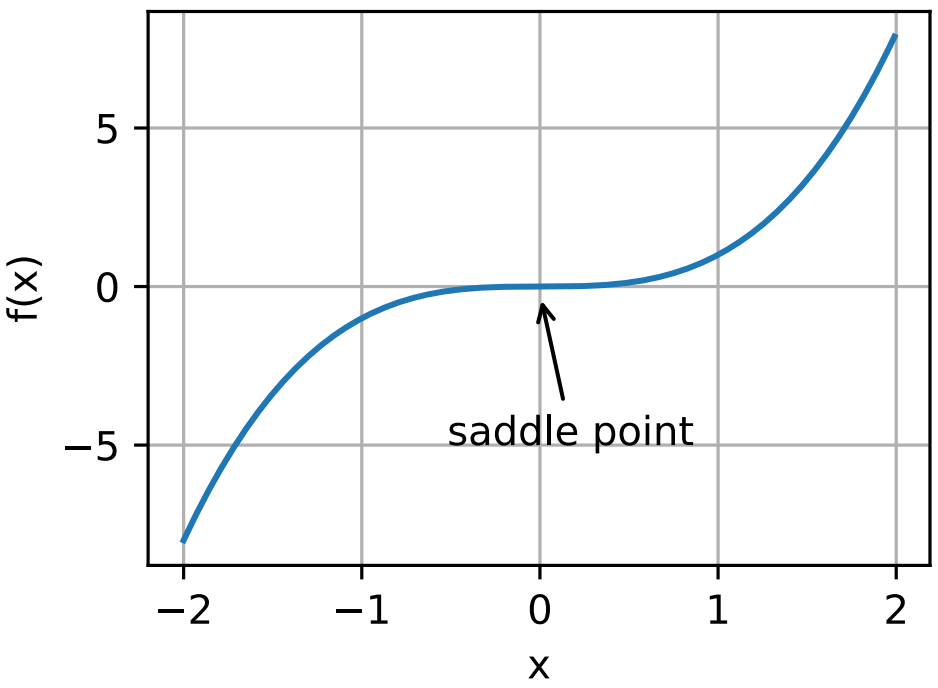

Saddle points in higher dimensions are even more insidious, as the example below shows. Consider the function $f(x, y) = x^2 - y^2$. It has its saddle point at $(0, 0)$. This is a maximum with respect to $y$ and a minimum with respect to $x$. Moreover, it *looks* like a saddle, which is where this mathematical property got its name.

```
x, y = torch.meshgrid(
    torch.linspace(-1.0, 1.0, 101), torch.linspace(-1.0, 1.0, 101))
z = x**2 - y**2

ax = d2l.plt.figure().add_subplot(111, projection='3d')
ax.plot_wireframe(x, y, z, **{'rstride': 10, 'cstride': 10})
```

```
ax.plot([0], [0], [0], 'rx')
ticks = [-1, 0, 1]
d2l.plt.xticks(ticks)
d2l.plt.yticks(ticks)
ax.set_zticks(ticks)
d2l.plt.xlabel('x')
d2l.plt.ylabel('y');
```

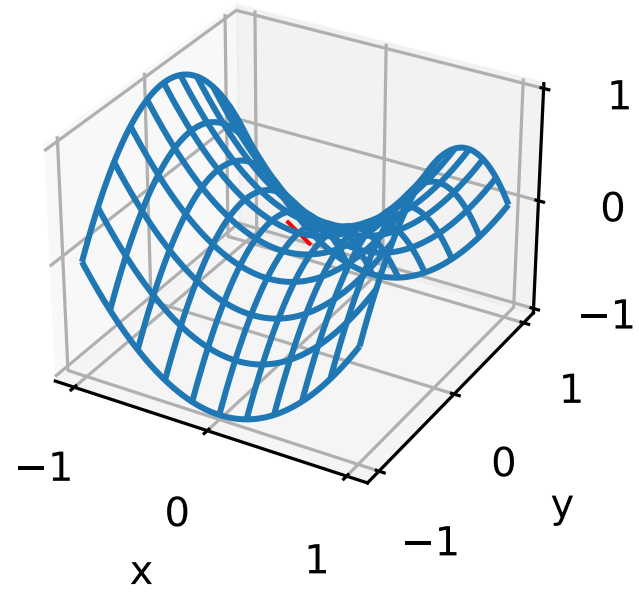

We assume that the input of a function is a $k$-dimensional vector and its output is a scalar, so its Hessian matrix will have $k$ eigenvalues. The solution of the function could be a local minimum, a local maximum, or a saddle point at a position where the function gradient is zero:

- When the eigenvalues of the function's Hessian matrix at the zero-gradient position are all positive, we have a local minimum for the function.
- When the eigenvalues of the function's Hessian matrix at the zero-gradient position are all negative, we have a local maximum for the function.
- When the eigenvalues of the function's Hessian matrix at the zero-gradient position are negative and positive, we have a saddle point for the function.

For high-dimensional problems the likelihood that at least *some* of the eigenvalues are negative is quite high. This makes saddle points more likely than local minima. We will discuss some exceptions to this situation in the next section when introducing convexity. In short, convex functions are those where the eigenvalues of the Hessian are never negative. Sadly, though, most deep learning problems do not fall into this category. Nonetheless it is a great tool to study optimization algorithms.

### Vanishing Gradients

Probably the most insidious problem to encounter is the vanishing gradient. Recall our commonly-used activation functions and their derivatives in Section 5.1.2. For instance, assume that we want to minimize the function $f(x) = \tanh(x)$ and we happen to get started at $x = 4$. As we can see, the gradient of $f$ is close to nil. More specifically, $f'(x) = 1 - \tanh^2(x)$ and thus $f'(4) = 0.0013$. Consequently, optimization will get stuck for a long time before we make progress. This turns out to be one of the reasons that training deep learning models was quite tricky prior to the introduction of the ReLU activation function.

```
x = torch.arange(-2.0, 5.0, 0.01)
d2l.plot(x, [torch.tanh(x)], 'x', 'f(x)')
annotate('vanishing gradient', (4, 1), (2, 0.0))
```

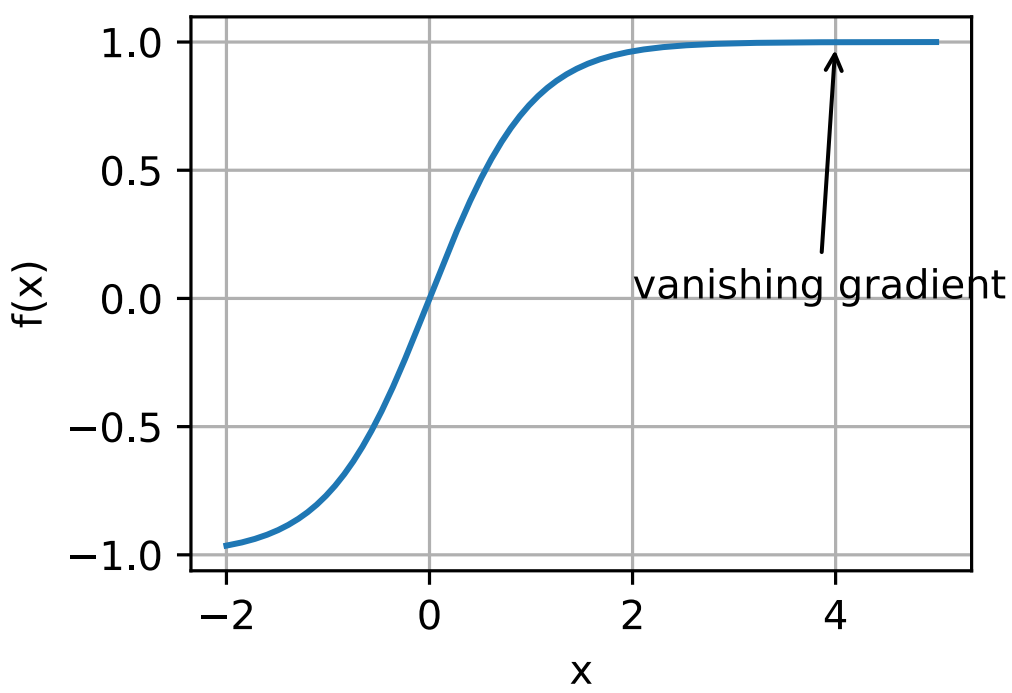

As we saw, optimization for deep learning is full of challenges. Fortunately there exists a robust range of algorithms that perform well and that are easy to use even for beginners. Furthermore, it is not really necessary to find *the* best solution. Local optima or even approximate solutions thereof are still very useful.

### 12.1.3 Summary

- Minimizing the training error does *not* guarantee that we find the best set of parameters to minimize the generalization error.
- The optimization problems may have many local minima.
- The problem may have even more saddle points, as generally the problems are not convex.
- Vanishing gradients can cause optimization to stall. Often a reparametrization of the problem helps. Good initialization of the parameters can be beneficial, too.

### 12.1.4 Exercises

1. Consider a simple MLP with a single hidden layer of, say, $d$ dimensions in the hidden layer and a single output. Show that for any local minimum there are at least $d!$ equivalent solutions that behave identically.
2. Assume that we have a symmetric random matrix $\mathbf{M}$ where the entries $M_{ij} = M_{ji}$ are each drawn from some probability distribution $p_{ij}$. Furthermore assume that $p_{ij}(x) = p_{ij}(-x)$, i.e., that the distribution is symmetric (see e.g., Wigner (1958) for details).

1. Prove that the distribution over eigenvalues is also symmetric. That is, for any eigenvector $\mathbf{v}$ the probability that the associated eigenvalue $\lambda$ satisfies $P(\lambda > 0) = P(\lambda < 0)$.
2. Why does the above *not* imply $P(\lambda > 0) = 0.5$?

3. What other challenges involved in deep learning optimization can you think of?
4. Assume that you want to balance a (real) ball on a (real) saddle.
   1. Why is this hard?
   2. Can you exploit this effect also for optimization algorithms?

Discussions[166].

## 12.2 Convexity

Convexity plays a vital role in the design of optimization algorithms. This is largely due to the fact that it is much easier to analyze and test algorithms in such a context. In other words, if the algorithm performs poorly even in the convex setting, typically we should not hope to see great results otherwise. Furthermore, even though the optimization problems in deep learning are generally nonconvex, they often exhibit some properties of convex ones near local minima. This can lead to exciting new optimization variants such as (Izmailov *et al.*, 2018).

```
%matplotlib inline
import numpy as np
import torch
from mpl_toolkits import mplot3d
from d2l import torch as d2l
```

### 12.2.1 Definitions

Before convex analysis, we need to define *convex sets* and *convex functions*. They lead to mathematical tools that are commonly applied to machine learning.

#### Convex Sets

Sets are the basis of convexity. Simply put, a set $\mathcal{X}$ in a vector space is *convex* if for any $a, b \in \mathcal{X}$ the line segment connecting $a$ and $b$ is also in $\mathcal{X}$. In mathematical terms this means that for all $\lambda \in [0, 1]$ we have

$$\lambda a + (1 - \lambda) b \in \mathcal{X} \text{ whenever } a, b \in \mathcal{X}. \tag{12.2.1}$$

This sounds a bit abstract. Consider Fig. 12.2.1. The first set is not convex since there exist line segments that are not contained in it. The other two sets suffer no such problem.

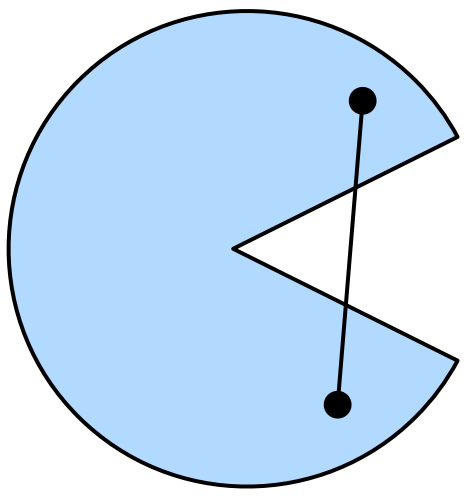
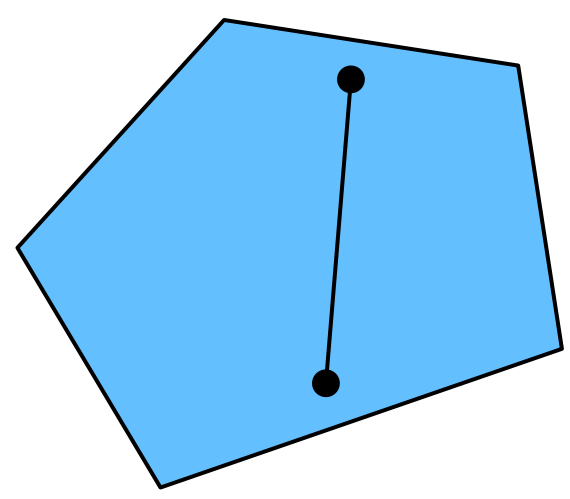
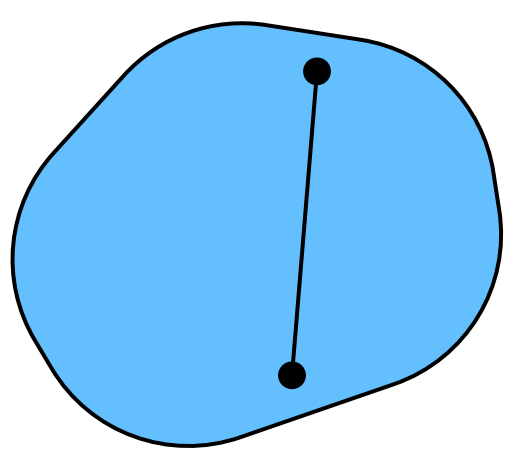

Fig. 12.2.1 The first set is nonconvex and the other two are convex.

Definitions on their own are not particularly useful unless you can do something with them. In this case we can look at intersections as shown in Fig. 12.2.2. Assume that $\mathcal{X}$ and $\mathcal{Y}$ are convex sets. Then $\mathcal{X} \cap \mathcal{Y}$ is also convex. To see this, consider any $a, b \in \mathcal{X} \cap \mathcal{Y}$. Since $\mathcal{X}$ and $\mathcal{Y}$ are convex, the line segments connecting $a$ and $b$ are contained in both $\mathcal{X}$ and $\mathcal{Y}$. Given that, they also need to be contained in $\mathcal{X} \cap \mathcal{Y}$, thus proving our theorem.

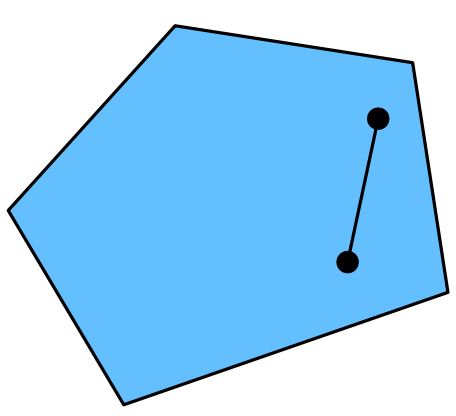
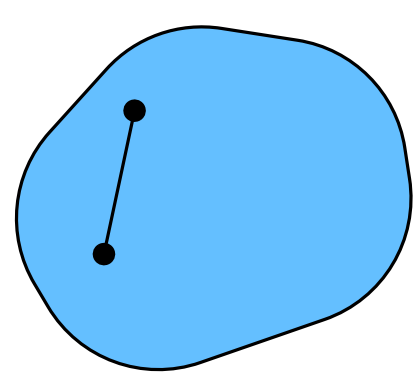
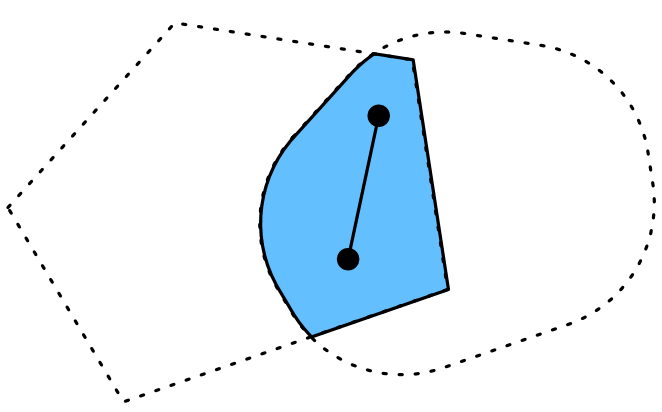

Fig. 12.2.2 The intersection between two convex sets is convex.

We can strengthen this result with little effort: given convex sets $\mathcal{X}_i$, their intersection $\cap_i \mathcal{X}_i$ is convex. To see that the converse is not true, consider two disjoint sets $\mathcal{X} \cap \mathcal{Y} = \emptyset$. Now pick $a \in \mathcal{X}$ and $b \in \mathcal{Y}$. The line segment in Fig. 12.2.3 connecting $a$ and $b$ needs to contain some part that is neither in $\mathcal{X}$ nor in $\mathcal{Y}$, since we assumed that $\mathcal{X} \cap \mathcal{Y} = \emptyset$. Hence the line segment is not in $\mathcal{X} \cup \mathcal{Y}$ either, thus proving that in general unions of convex sets need not be convex.

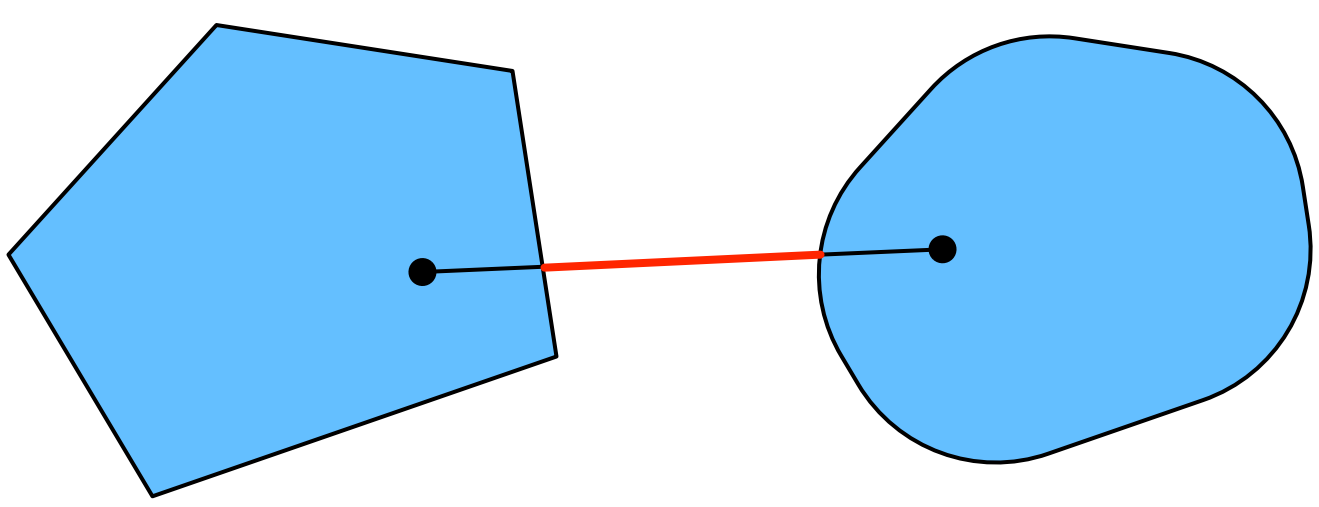

Fig. 12.2.3 The union of two convex sets need not be convex.

Typically the problems in deep learning are defined on convex sets. For instance, $\mathbb{R}^d$, the set of $d$-dimensional vectors of real numbers, is a convex set (after all, the line between any two points in $\mathbb{R}^d$ remains in $\mathbb{R}^d$). In some cases we work with variables of bounded length, such as balls of radius $r$ as defined by $\{\mathbf{x} | \mathbf{x} \in \mathbb{R}^d \text{ and } \|\mathbf{x}\| \leq r\}$.

### Convex Functions

Now that we have convex sets we can introduce *convex functions* $f$. Given a convex set $\mathcal{X}$, a function $f : \mathcal{X} \to \mathbb{R}$ is *convex* if for all $x, x' \in \mathcal{X}$ and for all $\lambda \in [0, 1]$ we have

$$\lambda f(x) + (1 - \lambda) f(x') \geq f(\lambda x + (1 - \lambda) x'). \tag{12.2.2}$$

To illustrate this let's plot a few functions and check which ones satisfy the requirement. Below we define a few functions, both convex and nonconvex.

```
f = lambda x: 0.5 * x**2  # Convex
g = lambda x: torch.cos(np.pi * x)  # Nonconvex
h = lambda x: torch.exp(0.5 * x)  # Convex

x, segment = torch.arange(-2, 2, 0.01), torch.tensor([-1.5, 1])
d2l.use_svg_display()
_, axes = d2l.plt.subplots(1, 3, figsize=(9, 3))
for ax, func in zip(axes, [f, g, h]):
    d2l.plot([x, segment], [func(x), func(segment)], axes=ax)
```

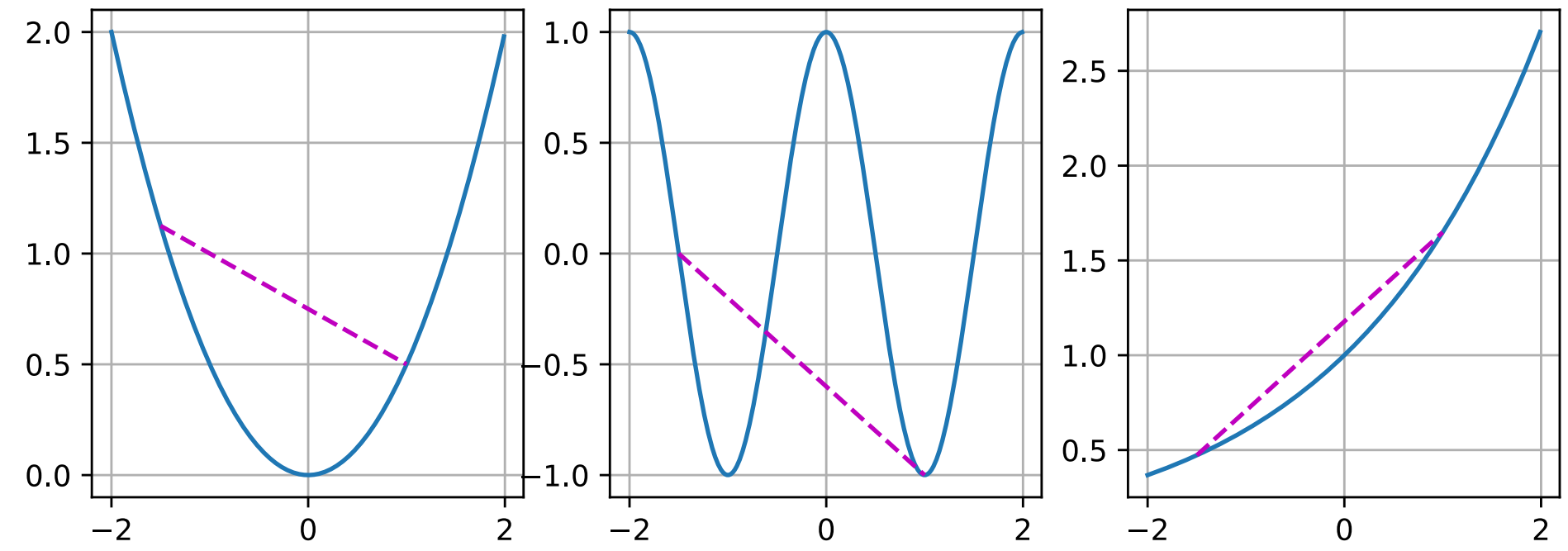

As expected, the cosine function is *nonconvex*, whereas the parabola and the exponential function are. Note that the requirement that $\mathcal{X}$ is a convex set is necessary for the condition to make sense. Otherwise the outcome of $f(\lambda x + (1 - \lambda) x')$ might not be well defined.

### Jensen's Inequality

Given a convex function $f$, one of the most useful mathematical tools is *Jensen's inequality*. It amounts to a generalization of the definition of convexity:

$$\sum_i \alpha_i f(x_i) \geq f\left(\sum_i \alpha_i x_i\right) \text{ and } E_X[f(X)] \geq f\left(E_X[X]\right), \tag{12.2.3}$$

where $\alpha_i$ are nonnegative real numbers such that $\sum_i \alpha_i = 1$ and $X$ is a random variable. In other words, the expectation of a convex function is no less than the convex function of an expectation, where the latter is usually a simpler expression. To prove the first inequality we repeatedly apply the definition of convexity to one term in the sum at a time.

One of the common applications of Jensen's inequality is to bound a more complicated expression by a simpler one. For example, its application can be with regard to the log-likelihood of partially observed random variables. That is, we use

$$E_{Y\sim P(Y)}[-\log P(X \mid Y)] \geq -\log P(X), \tag{12.2.4}$$

since $\int P(Y)P(X \mid Y)dY = P(X)$. This can be used in variational methods. Here $Y$ is typically the unobserved random variable, $P(Y)$ is the best guess of how it might be distributed, and $P(X)$ is the distribution with $Y$ integrated out. For instance, in clustering $Y$ might be the cluster labels and $P(X \mid Y)$ is the generative model when applying cluster labels.

## 12.2.2 Properties

Convex functions have many useful properties. We describe a few commonly-used ones below.

### Local Minima Are Global Minima

First and foremost, the local minima of convex functions are also the global minima. We can prove it by contradiction as follows.

Consider a convex function $f$ defined on a convex set $\mathcal{X}$. Suppose that $x^* \in \mathcal{X}$ is a local minimum: there exists a small positive value $p$ so that for $x \in \mathcal{X}$ that satisfies $0 < |x - x^*| \leq p$ we have $f(x^*) < f(x)$.

Assume that the local minimum $x^*$ is not the global minimum of $f$: there exists $x' \in \mathcal{X}$ for which $f(x') < f(x^*)$. There also exists $\lambda \in [0, 1)$ such as $\lambda = 1 - \frac{p}{|x^* - x'|}$ so that $0 < |\lambda x^* + (1 - \lambda)x' - x^*| \leq p$.

However, according to the definition of convex functions, we have

$$\begin{aligned}
f(\lambda x^* + (1 - \lambda)x') &\leq \lambda f(x^*) + (1 - \lambda)f(x') \\
&< \lambda f(x^*) + (1 - \lambda)f(x^*) \\
&= f(x^*),
\end{aligned} \tag{12.2.5}$$

which contradicts with our statement that $x^*$ is a local minimum. Therefore, there does not exist $x' \in \mathcal{X}$ for which $f(x') < f(x^*)$. The local minimum $x^*$ is also the global minimum.

For instance, the convex function $f(x) = (x - 1)^2$ has a local minimum at $x = 1$, which is also the global minimum.

```
f = lambda x: (x - 1) ** 2
d2l.set_figsize()
d2l.plot([x, segment], [f(x), f(segment)], 'x', 'f(x)')
```

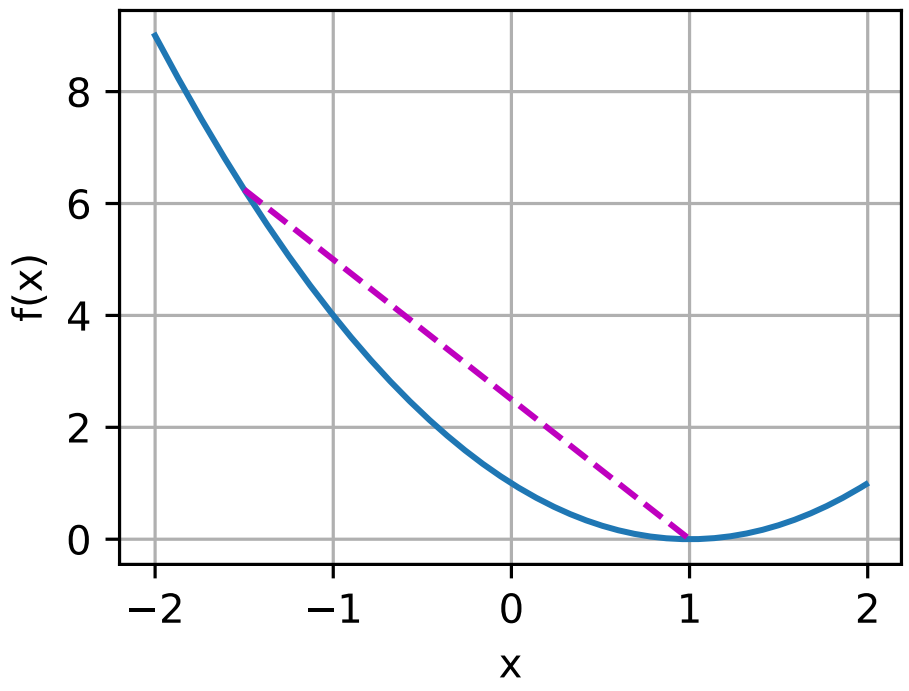

The fact that the local minima for convex functions are also the global minima is very convenient. It means that if we minimize functions we cannot "get stuck". Note, though, that this does not mean that there cannot be more than one global minimum or that there might even exist one. For instance, the function $f(x) = \max(|x| - 1, 0)$ attains its minimum value over the interval $[-1, 1]$. Conversely, the function $f(x) = \exp(x)$ does not attain a minimum value on $\mathbb{R}$: for $x \to -\infty$ it asymptotes to $0$, but there is no $x$ for which $f(x) = 0$.

### Below Sets of Convex Functions Are Convex

We can conveniently define convex sets via *below sets* of convex functions. Concretely, given a convex function $f$ defined on a convex set $\mathcal{X}$, any below set

$$\mathcal{S}_b \stackrel{\textrm{def}}{=} \{x | x \in \mathcal{X} \textrm{ and } f(x) \leq b\} \tag{12.2.6}$$

is convex.

Let's prove this quickly. Recall that for any $x, x' \in \mathcal{S}_b$ we need to show that $\lambda x + (1-\lambda) x' \in \mathcal{S}_b$ as long as $\lambda \in [0, 1]$. Since $f(x) \leq b$ and $f(x') \leq b$, by the definition of convexity we have

$$f(\lambda x + (1-\lambda) x') \leq \lambda f(x) + (1-\lambda) f(x') \leq b. \tag{12.2.7}$$

### Convexity and Second Derivatives

Whenever the second derivative of a function $f: \mathbb{R}^n \rightarrow \mathbb{R}$ exists it is very easy to check whether $f$ is convex. All we need to do is check whether the Hessian of $f$ is positive semidefinite: $\nabla^2f \succeq 0$, i.e., denoting the Hessian matrix $\nabla^2f$ by $\mathbf{H}$, $\mathbf{x}^\top \mathbf{H} \mathbf{x} \geq 0$ for all $\mathbf{x} \in \mathbb{R}^n$. For instance, the function $f(\mathbf{x}) = \frac{1}{2}\|\mathbf{x}\|^2$ is convex since $\nabla^2 f = \mathbf{1}$, i.e., its Hessian is an identity matrix.

Formally, a twice-differentiable one-dimensional function $f : \mathbb{R} \to \mathbb{R}$ is convex if and only if its second derivative $f'' \geq 0$. For any twice-differentiable multidimensional function $f : \mathbb{R}^n \to \mathbb{R}$, it is convex if and only if its Hessian $\nabla^2 f \succeq 0$.

First, we need to prove the one-dimensional case. To see that convexity of $f$ implies $f'' \geq 0$ we use the fact that

$$\frac{1}{2} f(x+\epsilon) + \frac{1}{2} f(x-\epsilon) \geq f\left(\frac{x+\epsilon}{2} + \frac{x-\epsilon}{2}\right) = f(x). \tag{12.2.8}$$

Since the second derivative is given by the limit over finite differences it follows that

$$f''(x) = \lim_{\epsilon \to 0} \frac{f(x+\epsilon) + f(x-\epsilon) - 2f(x)}{\epsilon^2} \geq 0. \tag{12.2.9}$$

To see that $f'' \geq 0$ implies that $f$ is convex we use the fact that $f'' \geq 0$ implies that $f'$ is a monotonically nondecreasing function. Let $a < x < b$ be three points in $\mathbb{R}$, where $x = (1-\lambda)a + \lambda b$ and $\lambda \in (0, 1)$. According to the mean value theorem, there exist $\alpha \in [a, x]$ and $\beta \in [x, b]$ such that

$$f'(\alpha) = \frac{f(x) - f(a)}{x - a} \text{ and } f'(\beta) = \frac{f(b) - f(x)}{b - x}. \tag{12.2.10}$$

By monotonicity $f'(\beta) \geq f'(\alpha)$, hence

$$\frac{x-a}{b-a} f(b) + \frac{b-x}{b-a} f(a) \geq f(x). \tag{12.2.11}$$

Since $x = (1-\lambda)a + \lambda b$, we have

$$\lambda f(b) + (1-\lambda) f(a) \geq f((1-\lambda)a + \lambda b), \tag{12.2.12}$$

thus proving convexity.

Second, we need a lemma before proving the multidimensional case: $f : \mathbb{R}^n \to \mathbb{R}$ is convex if and only if for all $\mathbf{x}, \mathbf{y} \in \mathbb{R}^n$

$$g(z) \stackrel{\text{def}}{=} f(z\mathbf{x} + (1-z)\mathbf{y}) \text{ where } z \in [0, 1] \tag{12.2.13}$$

is convex.

To prove that convexity of $f$ implies that $g$ is convex, we can show that for all $a, b, \lambda \in [0, 1]$ (thus $0 \leq \lambda a + (1-\lambda)b \leq 1$)

$$\begin{aligned}
&g(\lambda a + (1-\lambda)b)\\
=&f\left(\left(\lambda a + (1-\lambda)b\right)\mathbf{x} + \left(1 - \lambda a - (1-\lambda)b\right)\mathbf{y}\right)\\
=&f\left(\lambda\left(a\mathbf{x} + (1-a)\mathbf{y}\right) + (1-\lambda)\left(b\mathbf{x} + (1-b)\mathbf{y}\right)\right)\\
\leq&\lambda f\left(a\mathbf{x} + (1-a)\mathbf{y}\right) + (1-\lambda) f\left(b\mathbf{x} + (1-b)\mathbf{y}\right)\\
=&\lambda g(a) + (1-\lambda) g(b).
\end{aligned} \tag{12.2.14}$$

To prove the converse, we can show that for all $\lambda \in [0, 1]$

$$\begin{aligned}
& f(\lambda \mathbf{x} + (1-\lambda)\mathbf{y}) \\
=& g(\lambda \cdot 1 + (1-\lambda) \cdot 0) \\
\leq& \lambda g(1) + (1-\lambda) g(0) \\
=& \lambda f(\mathbf{x}) + (1-\lambda) f(\mathbf{y}).
\end{aligned} \tag{12.2.15}$$

Finally, using the lemma above and the result of the one-dimensional case, the multidimensional case can be proven as follows. A multidimensional function $f : \mathbb{R}^n \rightarrow \mathbb{R}$ is convex if and only if for all $\mathbf{x}, \mathbf{y} \in \mathbb{R}^n$ $g(z) \stackrel{\mathrm{def}}{=} f(z\mathbf{x} + (1-z)\mathbf{y})$, where $z \in [0, 1]$, is convex. According to the one-dimensional case, this holds if and only if $g'' = (\mathbf{x} - \mathbf{y})^\top \mathbf{H}(\mathbf{x} - \mathbf{y}) \geq 0$ ($\mathbf{H} \stackrel{\mathrm{def}}{=} \nabla^2 f$) for all $\mathbf{x}, \mathbf{y} \in \mathbb{R}^n$, which is equivalent to $\mathbf{H} \succeq 0$ per the definition of positive semidefinite matrices.

## 12.2.3 Constraints

One of the nice properties of convex optimization is that it allows us to handle constraints efficiently. That is, it allows us to solve *constrained optimization* problems of the form:

$$\begin{aligned}
& \underset{\mathbf{x}}{\mathrm{minimize}} \ f(\mathbf{x}) \\
& \text{subject to } c_i(\mathbf{x}) \leq 0 \text{ for all } i \in \{1, \ldots, n\},
\end{aligned} \tag{12.2.16}$$

where $f$ is the objective and the functions $c_i$ are constraint functions. To see what this does consider the case where $c_1(\mathbf{x}) = \|\mathbf{x}\|_2 - 1$. In this case the parameters $\mathbf{x}$ are constrained to the unit ball. If a second constraint is $c_2(\mathbf{x}) = \mathbf{v}^\top \mathbf{x} + b$, then this corresponds to all $\mathbf{x}$ lying on a half-space. Satisfying both constraints simultaneously amounts to selecting a slice of a ball.

### Lagrangian

In general, solving a constrained optimization problem is difficult. One way of addressing it stems from physics with a rather simple intuition. Imagine a ball inside a box. The ball will roll to the place that is lowest and the forces of gravity will be balanced out with the forces that the sides of the box can impose on the ball. In short, the gradient of the objective function (i.e., gravity) will be offset by the gradient of the constraint function (the ball need to remain inside the box by virtue of the walls "pushing back"). Note that some constraints may not be active: the walls that are not touched by the ball will not be able to exert any force on the ball.

Skipping over the derivation of the *Lagrangian* $L$, the above reasoning can be expressed via the following saddle point optimization problem:

$$L(\mathbf{x}, \alpha_1, \ldots, \alpha_n) = f(\mathbf{x}) + \sum_{i=1}^{n} \alpha_i c_i(\mathbf{x}) \text{ where } \alpha_i \geq 0. \tag{12.2.17}$$

Here the variables $\alpha_i$ ($i = 1, \ldots, n$) are the so-called *Lagrange multipliers* that ensure that constraints are properly enforced. They are chosen just large enough to ensure that $c_i(\mathbf{x}) \leq 0$ for all $i$. For instance, for any $\mathbf{x}$ where $c_i(\mathbf{x}) < 0$ naturally, we'd end up picking $\alpha_i = 0$. Moreover, this is a saddle point optimization problem where one wants to *maximize* $L$ with respect to all $\alpha_i$ and simultaneously *minimize* it with respect to $\mathbf{x}$. There is a rich body of literature explaining how to arrive at the function $L(\mathbf{x}, \alpha_1, \ldots, \alpha_n)$. For our purposes it is sufficient to know that the saddle point of $L$ is where the original constrained optimization problem is solved optimally.

## Penalties

One way of satisfying constrained optimization problems at least *approximately* is to adapt the Lagrangian $L$. Rather than satisfying $c_i(\mathbf{x}) \leq 0$ we simply add $\alpha_i c_i(\mathbf{x})$ to the objective function $f(x)$. This ensures that the constraints will not be violated too badly.

In fact, we have been using this trick all along. Consider weight decay in Section 3.7. In it we add $\frac{\lambda}{2}\|\mathbf{w}\|^2$ to the objective function to ensure that $\mathbf{w}$ does not grow too large. From the constrained optimization point of view we can see that this will ensure that $\|\mathbf{w}\|^2 - r^2 \leq 0$ for some radius $r$. Adjusting the value of $\lambda$ allows us to vary the size of $\mathbf{w}$.

In general, adding penalties is a good way of ensuring approximate constraint satisfaction. In practice this turns out to be much more robust than exact satisfaction. Furthermore, for nonconvex problems many of the properties that make the exact approach so appealing in the convex case (e.g., optimality) no longer hold.

## Projections

An alternative strategy for satisfying constraints is projections. Again, we encountered them before, e.g., when dealing with gradient clipping in Section 9.5. There we ensured that a gradient has length bounded by $\theta$ via

$$\mathbf{g} \leftarrow \mathbf{g} \cdot \mathrm{min}(1, \theta/\|\mathbf{g}\|). \tag{12.2.18}$$

This turns out to be a *projection* of $\mathbf{g}$ onto the ball of radius $\theta$. More generally, a projection on a convex set $\mathcal{X}$ is defined as

$$\mathrm{Proj}_\mathcal{X}(\mathbf{x}) = \mathop{\mathrm{argmin}}_{\mathbf{x}' \in \mathcal{X}} \|\mathbf{x} - \mathbf{x}'\|, \tag{12.2.19}$$

which is the closest point in $\mathcal{X}$ to $\mathbf{x}$.

The mathematical definition of projections may sound a bit abstract. Fig. 12.2.4 explains it somewhat more clearly. In it we have two convex sets, a circle and a diamond. Points inside both sets (yellow) remain unchanged during projections. Points outside both sets (black) are projected to the points inside the sets (red) that are closet to the original points (black). While for $\ell_2$ balls this leaves the direction unchanged, this need not be the case in general, as can be seen in the case of the diamond.

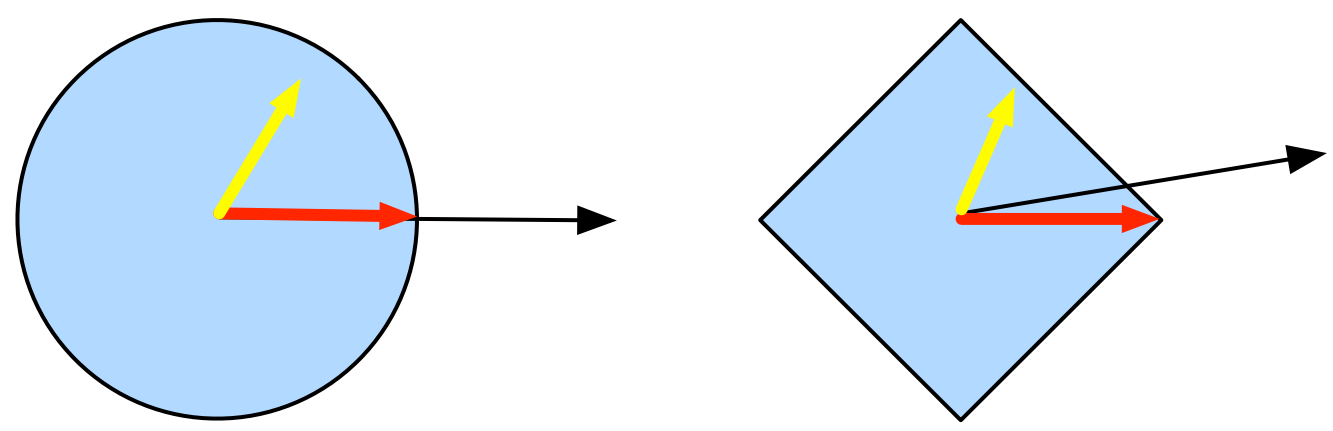

Fig. 12.2.4 Convex Projections.

One of the uses for convex projections is to compute sparse weight vectors. In this case we project weight vectors onto an $\ell_1$ ball, which is a generalized version of the diamond case in Fig. 12.2.4.

## 12.2.4 Summary

In the context of deep learning the main purpose of convex functions is to motivate optimization algorithms and help us understand them in detail. In the following we will see how gradient descent and stochastic gradient descent can be derived accordingly.

- Intersections of convex sets are convex. Unions are not.
- The expectation of a convex function is no less than the convex function of an expectation (Jensen's inequality).
- A twice-differentiable function is convex if and only if its Hessian (a matrix of second derivatives) is positive semidefinite.
- Convex constraints can be added via the Lagrangian. In practice we may simply add them with a penalty to the objective function.
- Projections map to points in the convex set closest to the original points.

## 12.2.5 Exercises

1. Assume that we want to verify convexity of a set by drawing all lines between points within the set and checking whether the lines are contained.
   1. Prove that it is sufficient to check only the points on the boundary.
   2. Prove that it is sufficient to check only the vertices of the set.
2. Denote by $\mathcal{B}_p[r] \stackrel{\textrm{def}}{=} \{\mathbf{x} | \mathbf{x} \in \mathbb{R}^d \textrm{ and } \|\mathbf{x}\|_p \leq r\}$ the ball of radius $r$ using the $p$-norm. Prove that $\mathcal{B}_p[r]$ is convex for all $p \geq 1$.
3. Given convex functions $f$ and $g$, show that $\mathrm{max}(f, g)$ is convex, too. Prove that $\mathrm{min}(f, g)$ is not convex.
4. Prove that the normalization of the softmax function is convex. More specifically prove the convexity of $f(x) = \log \sum_i \exp(x_i)$.

5. Prove that linear subspaces, i.e., $\mathcal{X} = \{\mathbf{x} | \mathbf{W}\mathbf{x} = \mathbf{b}\}$, are convex sets.
6. Prove that in the case of linear subspaces with $\mathbf{b} = \mathbf{0}$ the projection $\mathrm{Proj}_\mathcal{X}$ can be written as $\mathbf{M}\mathbf{x}$ for some matrix $\mathbf{M}$.
7. Show that for twice-differentiable convex functions $f$ we can write $f(x + \epsilon) = f(x) + \epsilon f'(x) + \frac{1}{2}\epsilon^2 f''(x + \xi)$ for some $\xi \in [0, \epsilon]$.
8. Given a convex set $\mathcal{X}$ and two vectors $\mathbf{x}$ and $\mathbf{y}$, prove that projections never increase distances, i.e., $\|\mathbf{x} - \mathbf{y}\| \geq \|\mathrm{Proj}_\mathcal{X}(\mathbf{x}) - \mathrm{Proj}_\mathcal{X}(\mathbf{y})\|$.

Discussions[167].

# 12.3 Gradient Descent

In this section we are going to introduce the basic concepts underlying *gradient descent*. Although it is rarely used directly in deep learning, an understanding of gradient descent is key to understanding stochastic gradient descent algorithms. For instance, the optimization problem might diverge due to an overly large learning rate. This phenomenon can already be seen in gradient descent. Likewise, preconditioning is a common technique in gradient descent and carries over to more advanced algorithms. Let's start with a simple special case.

## 12.3.1 One-Dimensional Gradient Descent

Gradient descent in one dimension is an excellent example to explain why the gradient descent algorithm may reduce the value of the objective function. Consider some continuously differentiable real-valued function $f: \mathbb{R} \rightarrow \mathbb{R}$. Using a Taylor expansion we obtain

$$f(x + \epsilon) = f(x) + \epsilon f'(x) + O(\epsilon^2). \tag{12.3.1}$$

That is, in first-order approximation $f(x+\epsilon)$ is given by the function value $f(x)$ and the first derivative $f'(x)$ at $x$. It is not unreasonable to assume that for small $\epsilon$ moving in the direction of the negative gradient will decrease $f$. To keep things simple we pick a fixed step size $\eta > 0$ and choose $\epsilon = -\eta f'(x)$. Plugging this into the Taylor expansion above we get

$$f(x - \eta f'(x)) = f(x) - \eta f'^2(x) + O(\eta^2 f'^2(x)). \tag{12.3.2}$$

If the derivative $f'(x) \neq 0$ does not vanish we make progress since $\eta f'^2(x) > 0$. Moreover, we can always choose $\eta$ small enough for the higher-order terms to become irrelevant. Hence we arrive at

$$f(x - \eta f'(x)) \lessapprox f(x). \tag{12.3.3}$$

This means that, if we use

$$x \leftarrow x - \eta f'(x) \tag{12.3.4}$$

to iterate $x$, the value of function $f(x)$ might decline. Therefore, in gradient descent we first choose an initial value $x$ and a constant $\eta > 0$ and then use them to continuously iterate $x$ until the stop condition is reached, for example, when the magnitude of the gradient $|f'(x)|$ is small enough or the number of iterations has reached a certain value.

For simplicity we choose the objective function $f(x) = x^2$ to illustrate how to implement gradient descent. Although we know that $x = 0$ is the solution to minimize $f(x)$, we still use this simple function to observe how $x$ changes.

```
%matplotlib inline
import numpy as np
import torch
from d2l import torch as d2l
```

```
def f(x):  # Objective function
    return x ** 2

def f_grad(x):  # Gradient (derivative) of the objective function
    return 2 * x
```

Next, we use $x = 10$ as the initial value and assume $\eta = 0.2$. Using gradient descent to iterate $x$ for 10 times we can see that, eventually, the value of $x$ approaches the optimal solution.

```
def gd(eta, f_grad):
    x = 10.0
    results = [x]
    for i in range(10):
        x -= eta * f_grad(x)
        results.append(float(x))
    print(f'epoch 10, x: {x:f}')
    return results

results = gd(0.2, f_grad)
```

```
epoch 10, x: 0.060466
```

The progress of optimizing over $x$ can be plotted as follows.

```
def show_trace(results, f):
    n = max(abs(min(results)), abs(max(results)))
    f_line = torch.arange(-n, n, 0.01)
    d2l.set_figsize()
    d2l.plot([f_line, results], [[f(x) for x in f_line], [
        f(x) for x in results]], 'x', 'f(x)', fmts=['-', '-o'])
```

```
show_trace(results, f)
```

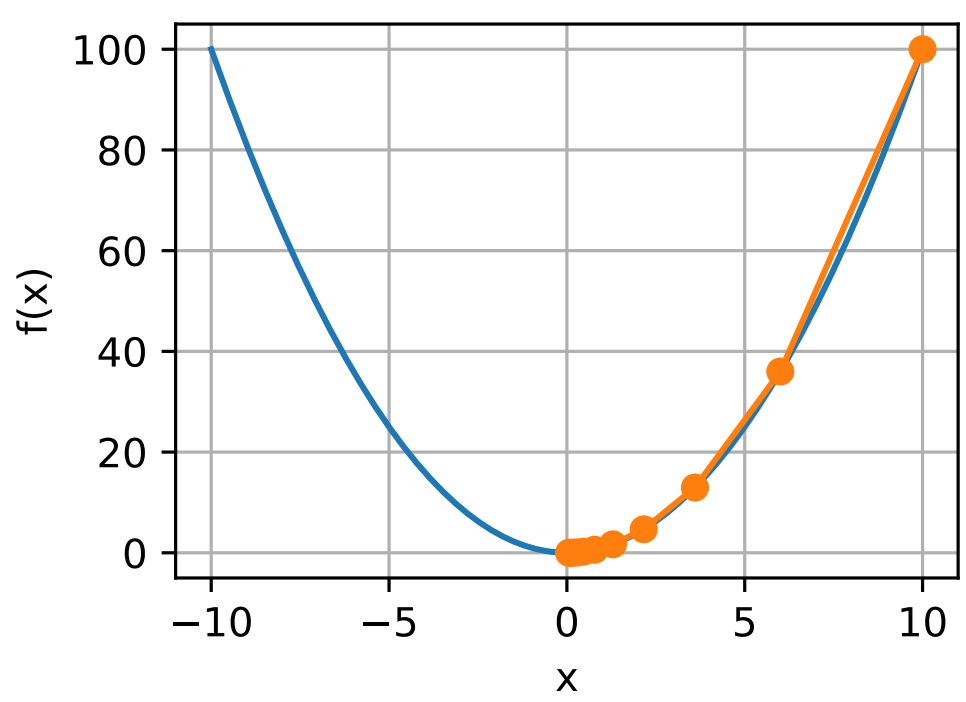

## Learning Rate

The learning rate $\eta$ can be set by the algorithm designer. If we use a learning rate that is too small, it will cause $x$ to update very slowly, requiring more iterations to get a better solution. To show what happens in such a case, consider the progress in the same optimization problem for $\eta = 0.05$. As we can see, even after 10 steps we are still very far from the optimal solution.

```
show_trace(gd(0.05, f_grad), f)
```

```
epoch 10, x: 3.486784
```

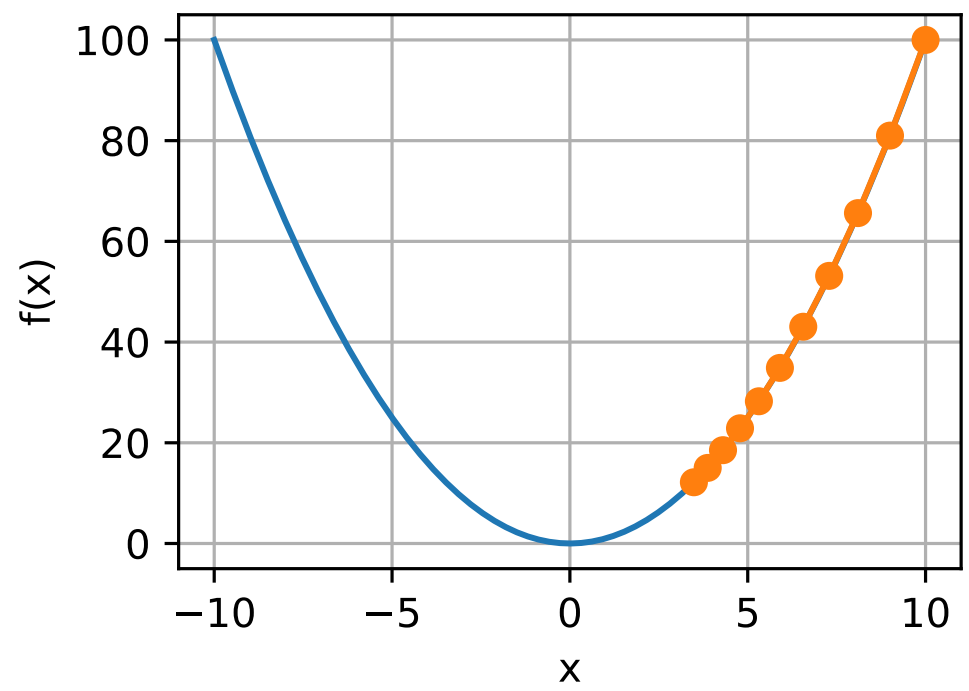

Conversely, if we use an excessively high learning rate, $\left|\eta f'(x)\right|$ might be too large for the first-order Taylor expansion formula. That is, the term $O(\eta^2 f'^2(x))$ in (12.3.2) might become significant. In this case, we cannot guarantee that the iteration of $x$ will be able to lower the value of $f(x)$. For example, when we set the learning rate to $\eta = 1.1$, $x$ overshoots the optimal solution $x = 0$ and gradually diverges.

```
show_trace(gd(1.1, f_grad), f)
```

```
epoch 10, x: 61.917364
```

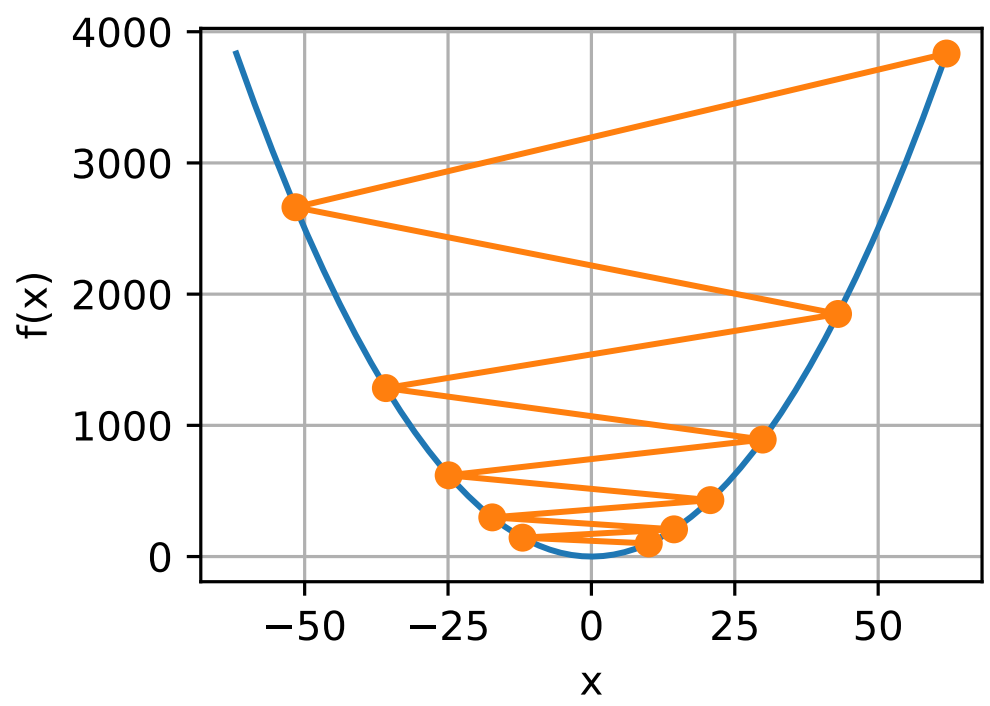

### Local Minima

To illustrate what happens for nonconvex functions consider the case of $f(x) = x \cdot \cos(cx)$ for some constant $c$. This function has infinitely many local minima. Depending on our choice of the learning rate and depending on how well conditioned the problem is, we may end up with one of many solutions. The example below illustrates how an (unrealistically) high learning rate will lead to a poor local minimum.

```
c = torch.tensor(0.15 * np.pi)

def f(x):  # Objective function
    return x * torch.cos(c * x)

def f_grad(x):  # Gradient of the objective function
    return torch.cos(c * x) - c * x * torch.sin(c * x)

show_trace(gd(2, f_grad), f)
```

```
epoch 10, x: -1.528166
```

## 12.3.2 Multivariate Gradient Descent

Now that we have a better intuition of the univariate case, let's consider the situation where $\mathbf{x} = [x_1, x_2, \ldots, x_d]^\top$. That is, the objective function $f : \mathbb{R}^d \to \mathbb{R}$ maps vectors into scalars.

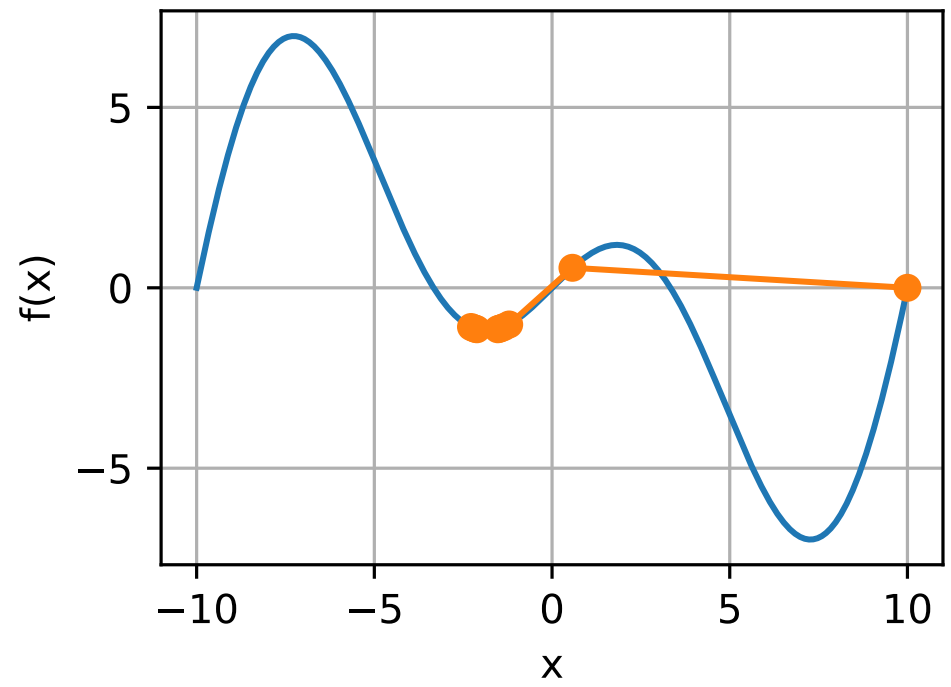

Correspondingly its gradient is multivariate, too. It is a vector consisting of $d$ partial derivatives:

$$\nabla f(\mathbf{x}) = \left[\frac{\partial f(\mathbf{x})}{\partial x_1}, \frac{\partial f(\mathbf{x})}{\partial x_2}, \ldots, \frac{\partial f(\mathbf{x})}{\partial x_d}\right]^\top. \tag{12.3.5}$$

Each partial derivative element $\partial f(\mathbf{x})/\partial x_i$ in the gradient indicates the rate of change of $f$ at $\mathbf{x}$ with respect to the input $x_i$. As before in the univariate case we can use the corresponding Taylor approximation for multivariate functions to get some idea of what we should do. In particular, we have that

$$f(\mathbf{x} + \boldsymbol{\epsilon}) = f(\mathbf{x}) + \boldsymbol{\epsilon}^\top \nabla f(\mathbf{x}) + O(\|\boldsymbol{\epsilon}\|^2). \tag{12.3.6}$$

In other words, up to second-order terms in $\boldsymbol{\epsilon}$ the direction of steepest descent is given by the negative gradient $-\nabla f(\mathbf{x})$. Choosing a suitable learning rate $\eta > 0$ yields the prototypical gradient descent algorithm:

$$\mathbf{x} \leftarrow \mathbf{x} - \eta \nabla f(\mathbf{x}). \tag{12.3.7}$$

To see how the algorithm behaves in practice let's construct an objective function $f(\mathbf{x}) = x_1^2 + 2x_2^2$ with a two-dimensional vector $\mathbf{x} = [x_1, x_2]^\top$ as input and a scalar as output. The gradient is given by $\nabla f(\mathbf{x}) = [2x_1, 4x_2]^\top$. We will observe the trajectory of $\mathbf{x}$ by gradient descent from the initial position $[-5, -2]$.

To begin with, we need two more helper functions. The first uses an update function and applies it 20 times to the initial value. The second helper visualizes the trajectory of $\mathbf{x}$.

```
def train_2d(trainer, steps=20, f_grad=None):  #@save
    """Optimize a 2D objective function with a customized trainer."""
    # `s1` and `s2` are internal state variables that will be used in Momentum,
↪ adagrad, RMSProp
    x1, x2, s1, s2 = -5, -2, 0, 0
    results = [(x1, x2)]
    for i in range(steps):
        if f_grad:
            x1, x2, s1, s2 = trainer(x1, x2, s1, s2, f_grad)
        else:
            x1, x2, s1, s2 = trainer(x1, x2, s1, s2)
```

(continued from previous page)

```
        results.append((x1, x2))
    print(f'epoch {i + 1}, x1: {float(x1):f}, x2: {float(x2):f}')
    return results
```

```
def show_trace_2d(f, results):  #@save
    """Show the trace of 2D variables during optimization."""
    d2l.set_figsize()
    d2l.plt.plot(*zip(*results), '-o', color='#ff7f0e')
    x1, x2 = torch.meshgrid(torch.arange(-5.5, 1.0, 0.1),
                            torch.arange(-3.0, 1.0, 0.1), indexing='ij')
    d2l.plt.contour(x1, x2, f(x1, x2), colors='#1f77b4')
    d2l.plt.xlabel('x1')
    d2l.plt.ylabel('x2')
```

Next, we observe the trajectory of the optimization variable $\mathbf{x}$ for learning rate $\eta = 0.1$. We can see that after 20 steps the value of $\mathbf{x}$ approaches its minimum at $[0, 0]$. Progress is fairly well-behaved albeit rather slow.

```
def f_2d(x1, x2):  # Objective function
    return x1 ** 2 + 2 * x2 ** 2

def f_2d_grad(x1, x2):  # Gradient of the objective function
    return (2 * x1, 4 * x2)

def gd_2d(x1, x2, s1, s2, f_grad):
    g1, g2 = f_grad(x1, x2)
    return (x1 - eta * g1, x2 - eta * g2, 0, 0)

eta = 0.1
show_trace_2d(f_2d, train_2d(gd_2d, f_grad=f_2d_grad))
```

```
epoch 20, x1: -0.057646, x2: -0.000073
```

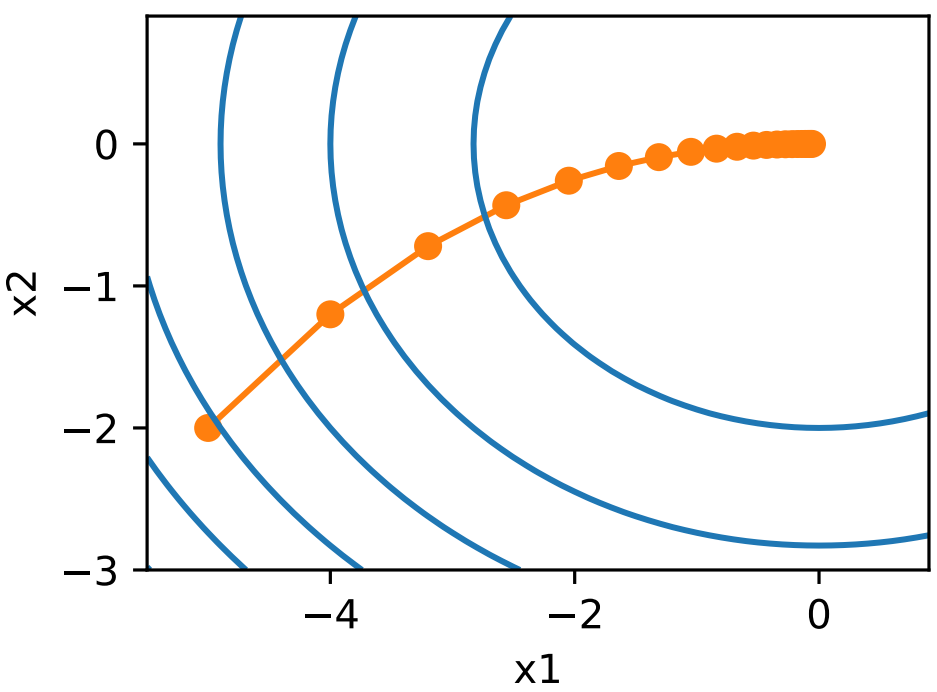

### 12.3.3 Adaptive Methods

As we could see in Section 12.3.1, getting the learning rate $\eta$ "just right" is tricky. If we pick it too small, we make little progress. If we pick it too large, the solution oscillates and in the worst case it might even diverge. What if we could determine $\eta$ automatically or get rid of having to select a learning rate at all? Second-order methods that look not only at the value and gradient of the objective function but also at its *curvature* can help in this case. While these methods cannot be applied to deep learning directly due to the computational cost, they provide useful intuition into how to design advanced optimization algorithms that mimic many of the desirable properties of the algorithms outlined below.

### Newton's Method

Reviewing the Taylor expansion of some function $f : \mathbb{R}^d \to \mathbb{R}$ there is no need to stop after the first term. In fact, we can write it as

$$f(\mathbf{x} + \boldsymbol{\epsilon}) = f(\mathbf{x}) + \boldsymbol{\epsilon}^\top \nabla f(\mathbf{x}) + \frac{1}{2}\boldsymbol{\epsilon}^\top \nabla^2 f(\mathbf{x}) \boldsymbol{\epsilon} + O(\|\boldsymbol{\epsilon}\|^3). \tag{12.3.8}$$

To avoid cumbersome notation we define $\mathbf{H} \stackrel{\mathrm{def}}{=} \nabla^2 f(\mathbf{x})$ to be the Hessian of $f$, which is a $d \times d$ matrix. For small $d$ and simple problems $\mathbf{H}$ is easy to compute. For deep neural networks, on the other hand, $\mathbf{H}$ may be prohibitively large, due to the cost of storing $O(d^2)$ entries. Furthermore it may be too expensive to compute via backpropagation. For now let's ignore such considerations and look at what algorithm we would get.

After all, the minimum of $f$ satisfies $\nabla f = 0$. Following calculus rules in Section 2.4.3, by taking derivatives of (12.3.8) with regard to $\boldsymbol{\epsilon}$ and ignoring higher-order terms we arrive at

$$\nabla f(\mathbf{x}) + \mathbf{H}\boldsymbol{\epsilon} = 0 \text{ and hence } \boldsymbol{\epsilon} = -\mathbf{H}^{-1} \nabla f(\mathbf{x}). \tag{12.3.9}$$

That is, we need to invert the Hessian $\mathbf{H}$ as part of the optimization problem.

As a simple example, for $f(x) = \frac{1}{2}x^2$ we have $\nabla f(x) = x$ and $\mathbf{H} = 1$. Hence for any $x$ we obtain $\epsilon = -x$. In other words, a *single* step is sufficient to converge perfectly without the need for any adjustment! Alas, we got a bit lucky here: the Taylor expansion was exact since $f(x+\epsilon) = \frac{1}{2}x^2 + \epsilon x + \frac{1}{2}\epsilon^2$.

Let's see what happens in other problems. Given a convex hyperbolic cosine function $f(x) = \cosh(cx)$ for some constant $c$, we can see that the global minimum at $x = 0$ is reached after a few iterations.

```
c = torch.tensor(0.5)

def f(x):  # Objective function
    return torch.cosh(c * x)

def f_grad(x):  # Gradient of the objective function
    return c * torch.sinh(c * x)
```

```
def f_hess(x):  # Hessian of the objective function
    return c**2 * torch.cosh(c * x)

def newton(eta=1):
    x = 10.0
    results = [x]
    for i in range(10):
        x -= eta * f_grad(x) / f_hess(x)
        results.append(float(x))
    print('epoch 10, x:', x)
    return results

show_trace(newton(), f)
```

```
epoch 10, x: tensor(0.)
```

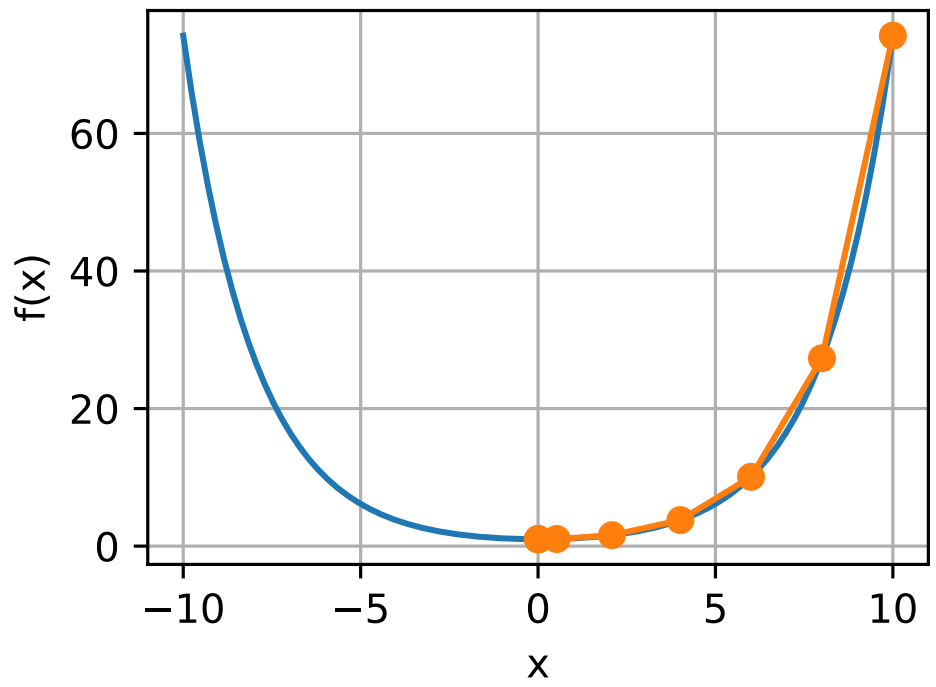

Now let's consider a *nonconvex* function, such as $f(x) = x\cos(cx)$ for some constant $c$. After all, note that in Newton's method we end up dividing by the Hessian. This means that if the second derivative is *negative* we may walk into the direction of *increasing* the value of $f$. That is a fatal flaw of the algorithm. Let's see what happens in practice.

```
c = torch.tensor(0.15 * np.pi)

def f(x):  # Objective function
    return x * torch.cos(c * x)

def f_grad(x):  # Gradient of the objective function
    return torch.cos(c * x) - c * x * torch.sin(c * x)

def f_hess(x):  # Hessian of the objective function
    return - 2 * c * torch.sin(c * x) - x * c**2 * torch.cos(c * x)

show_trace(newton(), f)
```

```
epoch 10, x: tensor(26.8341)
```

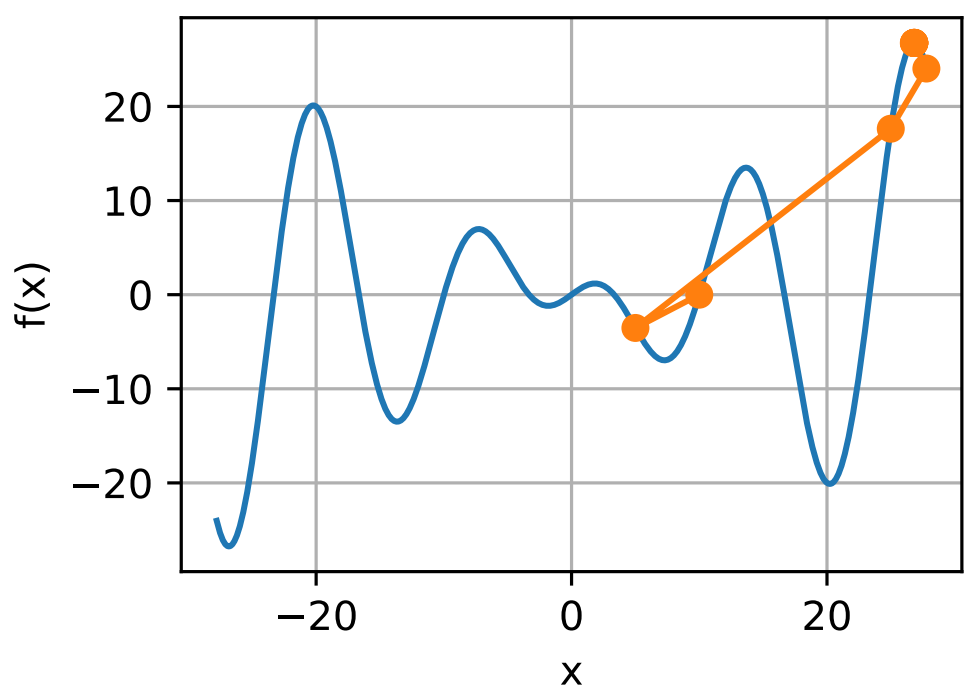

This went spectacularly wrong. How can we fix it? One way would be to "fix" the Hessian by taking its absolute value instead. Another strategy is to bring back the learning rate. This seems to defeat the purpose, but not quite. Having second-order information allows us to be cautious whenever the curvature is large and to take longer steps whenever the objective function is flatter. Let's see how this works with a slightly smaller learning rate, say $\eta = 0.5$. As we can see, we have quite an efficient algorithm.

```
show_trace(newton(0.5), f)
```

```
epoch 10, x: tensor(7.2699)
```

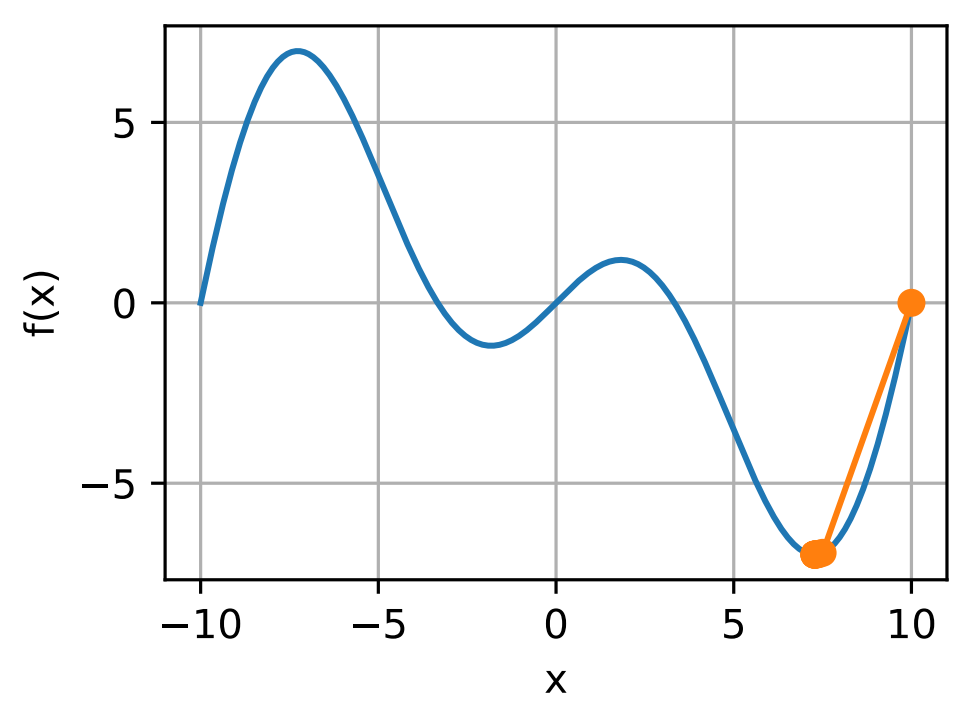

## Convergence Analysis

We only analyze the convergence rate of Newton's method for some convex and three times differentiable objective function $f$, where the second derivative is nonzero, i.e., $f'' > 0$. The multivariate proof is a straightforward extension of the one-dimensional argument below and omitted since it does not help us much in terms of intuition.

Denote by $x^{(k)}$ the value of $x$ at the $k^{\text{th}}$ iteration and let $e^{(k)} \stackrel{\text{def}}{=} x^{(k)} - x^*$ be the distance from optimality at the $k^{\text{th}}$ iteration. By Taylor expansion we have that the condition $f'(x^*) = 0$ can be written as

$$0 = f'(x^{(k)} - e^{(k)}) = f'(x^{(k)}) - e^{(k)} f''(x^{(k)}) + \frac{1}{2}(e^{(k)})^2 f'''(\xi^{(k)}), \tag{12.3.10}$$

which holds for some $\xi^{(k)} \in [x^{(k)} - e^{(k)}, x^{(k)}]$. Dividing the above expansion by $f''(x^{(k)})$ yields

$$e^{(k)} - \frac{f'(x^{(k)})}{f''(x^{(k)})} = \frac{1}{2}(e^{(k)})^2 \frac{f'''(\xi^{(k)})}{f''(x^{(k)})}. \tag{12.3.11}$$

Recall that we have the update $x^{(k+1)} = x^{(k)} - f'(x^{(k)})/f''(x^{(k)})$. Plugging in this update equation and taking the absolute value of both sides, we have

$$\left|e^{(k+1)}\right| = \frac{1}{2}(e^{(k)})^2 \frac{\left|f'''(\xi^{(k)})\right|}{f''(x^{(k)})}. \tag{12.3.12}$$

Consequently, whenever we are in a region of bounded $\left|f'''(\xi^{(k)})\right| / (2f''(x^{(k)})) \leq c$, we have a quadratically decreasing error

$$\left|e^{(k+1)}\right| \leq c(e^{(k)})^2. \tag{12.3.13}$$

As an aside, optimization researchers call this *linear* convergence, whereas a condition such as $\left|e^{(k+1)}\right| \leq \alpha \left|e^{(k)}\right|$ would be called a *constant* rate of convergence. Note that this analysis comes with a number of caveats. First, we do not really have much of a guarantee when we will reach the region of rapid convergence. Instead, we only know that once we reach it, convergence will be very quick. Second, this analysis requires that $f$ is well-behaved up to higher-order derivatives. It comes down to ensuring that $f$ does not have any "surprising" properties in terms of how it might change its values.

## Preconditioning

Quite unsurprisingly computing and storing the full Hessian is very expensive. It is thus desirable to find alternatives. One way to improve matters is *preconditioning*. It avoids computing the Hessian in its entirety but only computes the *diagonal* entries. This leads to update algorithms of the form

$$\mathbf{x} \leftarrow \mathbf{x} - \eta \text{diag}(\mathbf{H})^{-1} \nabla f(\mathbf{x}). \tag{12.3.14}$$

While this is not quite as good as the full Newton's method, it is still much better than not using it. To see why this might be a good idea consider a situation where one variable denotes height in millimeters and the other one denotes height in kilometers. Assuming that for both the natural scale is in meters, we have a terrible mismatch in parametrizations. Fortunately, using preconditioning removes this. Effectively preconditioning with gradient descent amounts to selecting a different learning rate for each variable (coordinate of vector $\mathbf{x}$). As we will see later, preconditioning drives some of the innovation in stochastic gradient descent optimization algorithms.

#### Gradient Descent with Line Search

One of the key problems in gradient descent is that we might overshoot the goal or make insufficient progress. A simple fix for the problem is to use line search in conjunction with gradient descent. That is, we use the direction given by $\nabla f(\mathbf{x})$ and then perform binary search as to which learning rate $\eta$ minimizes $f(\mathbf{x} - \eta\nabla f(\mathbf{x}))$.

This algorithm converges rapidly (for an analysis and proof see e.g., Boyd and Vandenberghe (2004)). However, for the purpose of deep learning this is not quite so feasible, since each step of the line search would require us to evaluate the objective function on the entire dataset. This is way too costly to accomplish.

### 12.3.4 Summary

- Learning rates matter. Too large and we diverge, too small and we do not make progress.
- Gradient descent can get stuck in local minima.
- In high dimensions adjusting the learning rate is complicated.
- Preconditioning can help with scale adjustment.
- Newton's method is a lot faster once it has started working properly in convex problems.
- Beware of using Newton's method without any adjustments for nonconvex problems.

### 12.3.5 Exercises

1. Experiment with different learning rates and objective functions for gradient descent.
2. Implement line search to minimize a convex function in the interval $[a, b]$.
    1. Do you need derivatives for binary search, i.e., to decide whether to pick $[a, (a+b)/2]$ or $[(a+b)/2, b]$.
    2. How rapid is the rate of convergence for the algorithm?
    3. Implement the algorithm and apply it to minimizing $\log(\exp(x) + \exp(-2x - 3))$.
3. Design an objective function defined on $\mathbb{R}^2$ where gradient descent is exceedingly slow. Hint: scale different coordinates differently.
4. Implement the lightweight version of Newton's method using preconditioning:
    1. Use diagonal Hessian as preconditioner.

2. Use the absolute values of that rather than the actual (possibly signed) values.
   3. Apply this to the problem above.
5. Apply the algorithm above to a number of objective functions (convex or not). What happens if you rotate coordinates by 45 degrees?

Discussions[168].

## 12.4 Stochastic Gradient Descent

In earlier chapters we kept using stochastic gradient descent in our training procedure, however, without explaining why it works. To shed some light on it, we just described the basic principles of gradient descent in Section 12.3. In this section, we go on to discuss *stochastic gradient descent* in greater detail.

```
%matplotlib inline
import math
import torch
from d2l import torch as d2l
```

### 12.4.1 Stochastic Gradient Updates

In deep learning, the objective function is usually the average of the loss functions for each example in the training dataset. Given a training dataset of $n$ examples, we assume that $f_i(\mathbf{x})$ is the loss function with respect to the training example of index $i$, where $\mathbf{x}$ is the parameter vector. Then we arrive at the objective function

$$f(\mathbf{x}) = \frac{1}{n} \sum_{i=1}^{n} f_i(\mathbf{x}). \tag{12.4.1}$$

The gradient of the objective function at $\mathbf{x}$ is computed as

$$\nabla f(\mathbf{x}) = \frac{1}{n} \sum_{i=1}^{n} \nabla f_i(\mathbf{x}). \tag{12.4.2}$$

If gradient descent is used, the computational cost for each independent variable iteration is $\mathcal{O}(n)$, which grows linearly with $n$. Therefore, when the training dataset is larger, the cost of gradient descent for each iteration will be higher.

Stochastic gradient descent (SGD) reduces computational cost at each iteration. At each iteration of stochastic gradient descent, we uniformly sample an index $i \in \{1, \ldots, n\}$ for data examples at random, and compute the gradient $\nabla f_i(\mathbf{x})$ to update $\mathbf{x}$:

$$\mathbf{x} \leftarrow \mathbf{x} - \eta \nabla f_i(\mathbf{x}), \tag{12.4.3}$$

where $\eta$ is the learning rate. We can see that the computational cost for each iteration drops from $O(n)$ of the gradient descent to the constant $O(1)$. Moreover, we want to emphasize that the stochastic gradient $\nabla f_i(\mathbf{x})$ is an unbiased estimate of the full gradient $\nabla f(\mathbf{x})$ because

$$\mathbb{E}_i \nabla f_i(\mathbf{x}) = \frac{1}{n} \sum_{i=1}^{n} \nabla f_i(\mathbf{x}) = \nabla f(\mathbf{x}). \tag{12.4.4}$$

This means that, on average, the stochastic gradient is a good estimate of the gradient.

Now, we will compare it with gradient descent by adding random noise with a mean of 0 and a variance of 1 to the gradient to simulate a stochastic gradient descent.

```
def f(x1, x2):  # Objective function
    return x1 ** 2 + 2 * x2 ** 2

def f_grad(x1, x2):  # Gradient of the objective function
    return 2 * x1, 4 * x2
```

```
def sgd(x1, x2, s1, s2, f_grad):
    g1, g2 = f_grad(x1, x2)
    # Simulate noisy gradient
    g1 += torch.normal(0.0, 1, (1,)).item()
    g2 += torch.normal(0.0, 1, (1,)).item()
    eta_t = eta * lr()
    return (x1 - eta_t * g1, x2 - eta_t * g2, 0, 0)
```

```
def constant_lr():
    return 1

eta = 0.1
lr = constant_lr  # Constant learning rate
d2l.show_trace_2d(f, d2l.train_2d(sgd, steps=50, f_grad=f_grad))
```

```
epoch 50, x1: -0.299320, x2: -0.024252
```

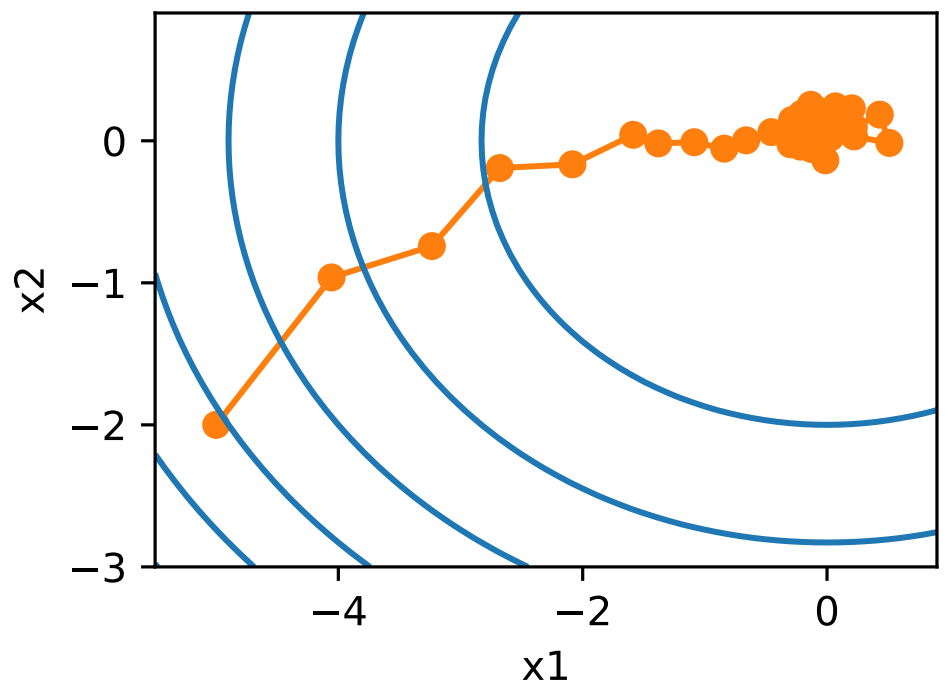

As we can see, the trajectory of the variables in the stochastic gradient descent is much more noisy than the one we observed in gradient descent in Section 12.3. This is due to the stochastic nature of the gradient. That is, even when we arrive near the minimum, we are still subject to the uncertainty injected by the instantaneous gradient via $\eta \nabla f_i(\mathbf{x})$. Even after 50 steps the quality is still not so good. Even worse, it will not improve after additional steps (we encourage you to experiment with a larger number of steps to confirm this). This leaves us with the only alternative: change the learning rate $\eta$. However, if we pick this too small, we will not make any meaningful progress initially. On the other hand, if we pick it too large, we will not get a good solution, as seen above. The only way to resolve these conflicting goals is to reduce the learning rate *dynamically* as optimization progresses.

This is also the reason for adding a learning rate function `lr` into the `sgd` step function. In the example above any functionality for learning rate scheduling lies dormant as we set the associated `lr` function to be constant.

## 12.4.2 Dynamic Learning Rate

Replacing $\eta$ with a time-dependent learning rate $\eta(t)$ adds to the complexity of controlling convergence of an optimization algorithm. In particular, we need to figure out how rapidly $\eta$ should decay. If it is too quick, we will stop optimizing prematurely. If we decrease it too slowly, we waste too much time on optimization. The following are a few basic strategies that are used in adjusting $\eta$ over time (we will discuss more advanced strategies later):

$$
\begin{aligned}
\eta(t) & = \eta_i \text{ if } t_i \leq t \leq t_{i+1} && \text{piecewise constant} \\
\eta(t) & = \eta_0 \cdot e^{-\lambda t} && \text{exponential decay} \\
\eta(t) & = \eta_0 \cdot (\beta t + 1)^{-\alpha} && \text{polynomial decay}
\end{aligned}
\tag{12.4.5}
$$

In the first *piecewise constant* scenario we decrease the learning rate, e.g., whenever progress in optimization stalls. This is a common strategy for training deep networks. Alternatively we could decrease it much more aggressively by an *exponential decay*. Unfortunately this often leads to premature stopping before the algorithm has converged. A popular choice is *polynomial decay* with $\alpha = 0.5$. In the case of convex optimization there are a number of proofs that show that this rate is well behaved.

Let's see what the exponential decay looks like in practice.

```python
def exponential_lr():
    # Global variable that is defined outside this function and updated inside
    global t
    t += 1
    return math.exp(-0.1 * t)

t = 1
lr = exponential_lr
d2l.show_trace_2d(f, d2l.train_2d(sgd, steps=1000, f_grad=f_grad))
```

```
epoch 1000, x1: -0.896277, x2: 0.012123
```

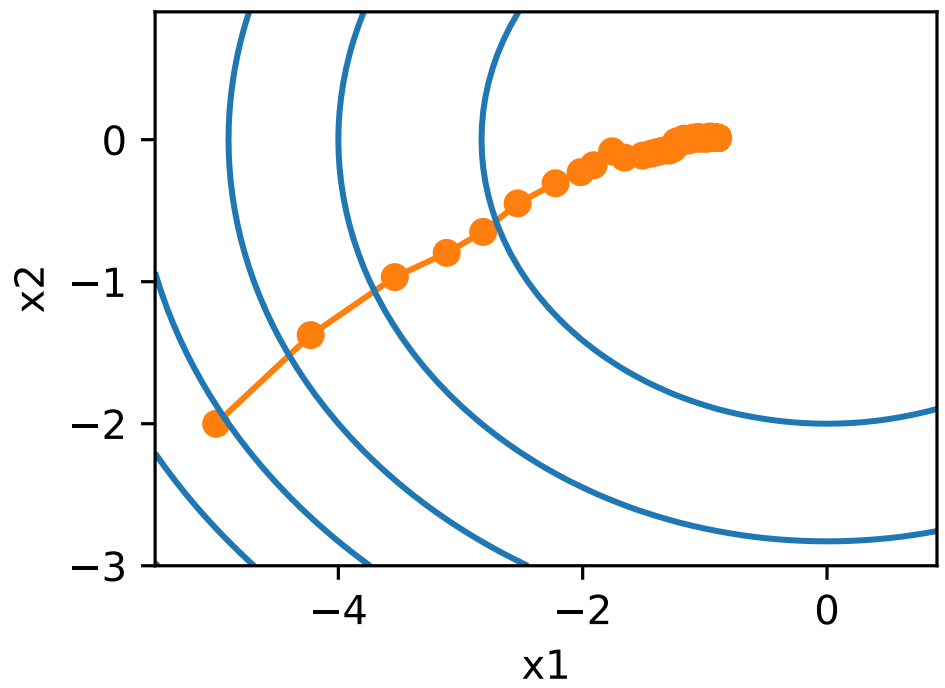

As expected, the variance in the parameters is significantly reduced. However, this comes at the expense of failing to converge to the optimal solution $\mathbf{x} = (0, 0)$. Even after 1000 iteration steps are we are still very far away from the optimal solution. Indeed, the algorithm fails to converge at all. On the other hand, if we use a polynomial decay where the learning rate decays with the inverse square root of the number of steps, convergence gets better after only 50 steps.

```
def polynomial_lr():
    # Global variable that is defined outside this function and updated inside
    global t
    t += 1
    return (1 + 0.1 * t) ** (-0.5)

t = 1
lr = polynomial_lr
d2l.show_trace_2d(f, d2l.train_2d(sgd, steps=50, f_grad=f_grad))
```

```
epoch 50, x1: 0.053842, x2: 0.059013
```

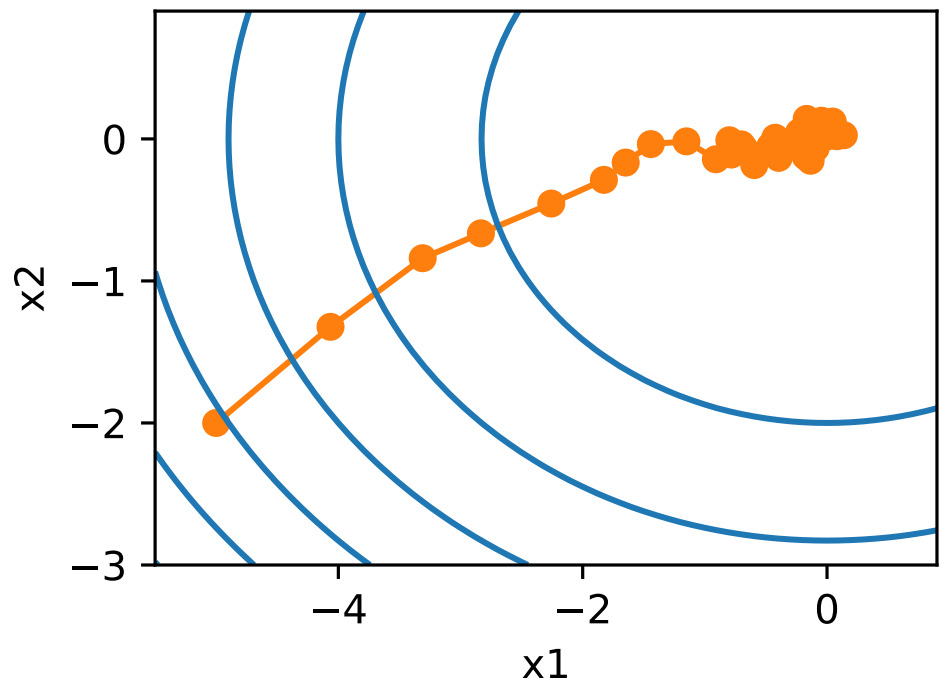

There exist many more choices for how to set the learning rate. For instance, we could start with a small rate, then rapidly ramp up and then decrease it again, albeit more slowly. We could even alternate between smaller and larger learning rates. There exists a large variety of such schedules. For now let's focus on learning rate schedules for which a comprehensive theoretical analysis is possible, i.e., on learning rates in a convex setting. For general nonconvex problems it is very difficult to obtain meaningful convergence guarantees, since in general minimizing nonlinear nonconvex problems is NP hard. For a survey see e.g., the excellent lecture notes[169] of Tibshirani 2015.

169

### 12.4.3 Convergence Analysis for Convex Objectives

The following convergence analysis of stochastic gradient descent for convex objective functions is optional and primarily serves to convey more intuition about the problem. We limit ourselves to one of the simplest proofs (Nesterov and Vial, 2000). Significantly more advanced proof techniques exist, e.g., whenever the objective function is particularly well behaved.

Suppose that the objective function $f(\boldsymbol{\xi}, \mathbf{x})$ is convex in $\mathbf{x}$ for all $\boldsymbol{\xi}$. More concretely, we consider the stochastic gradient descent update:

$$\mathbf{x}_{t+1} = \mathbf{x}_t - \eta_t \partial_{\mathbf{x}} f(\boldsymbol{\xi}_t, \mathbf{x}), \tag{12.4.6}$$

where $f(\boldsymbol{\xi}_t, \mathbf{x})$ is the objective function with respect to the training example $\boldsymbol{\xi}_t$ drawn from some distribution at step $t$ and $\mathbf{x}$ is the model parameter. Denote by

$$R(\mathbf{x}) = E_{\boldsymbol{\xi}}[f(\boldsymbol{\xi}, \mathbf{x})] \tag{12.4.7}$$

the expected risk and by $R^*$ its minimum with regard to $\mathbf{x}$. Last let $\mathbf{x}^*$ be the minimizer (we assume that it exists within the domain where $\mathbf{x}$ is defined). In this case we can track the distance between the current parameter $\mathbf{x}_t$ at time $t$ and the risk minimizer $\mathbf{x}^*$ and see whether it improves over time:

$$\begin{aligned}
&\|\mathbf{x}_{t+1} - \mathbf{x}^*\|^2 \\
=&\|\mathbf{x}_t - \eta_t \partial_{\mathbf{x}} f(\boldsymbol{\xi}_t, \mathbf{x}) - \mathbf{x}^*\|^2 \\
=&\|\mathbf{x}_t - \mathbf{x}^*\|^2 + \eta_t^2 \|\partial_{\mathbf{x}} f(\boldsymbol{\xi}_t, \mathbf{x})\|^2 - 2\eta_t \left\langle \mathbf{x}_t - \mathbf{x}^*, \partial_{\mathbf{x}} f(\boldsymbol{\xi}_t, \mathbf{x})\right\rangle.
\end{aligned} \tag{12.4.8}$$

We assume that the $\ell_2$ norm of stochastic gradient $\partial_{\mathbf{x}} f(\boldsymbol{\xi}_t, \mathbf{x})$ is bounded by some constant $L$, hence we have that

$$\eta_t^2 \|\partial_{\mathbf{x}} f(\boldsymbol{\xi}_t, \mathbf{x})\|^2 \leq \eta_t^2 L^2. \tag{12.4.9}$$

We are mostly interested in how the distance between $\mathbf{x}_t$ and $\mathbf{x}^*$ changes *in expectation*. In fact, for any specific sequence of steps the distance might well increase, depending on whichever $\boldsymbol{\xi}_t$ we encounter. Hence we need to bound the dot product. Since for any convex function $f$ it holds that $f(\mathbf{y}) \geq f(\mathbf{x}) + \langle f'(\mathbf{x}), \mathbf{y} - \mathbf{x}\rangle$ for all $\mathbf{x}$ and $\mathbf{y}$, by convexity we have

$$f(\boldsymbol{\xi}_t, \mathbf{x}^*) \geq f(\boldsymbol{\xi}_t, \mathbf{x}_t) + \left\langle \mathbf{x}^* - \mathbf{x}_t, \partial_{\mathbf{x}} f(\boldsymbol{\xi}_t, \mathbf{x}_t)\right\rangle. \tag{12.4.10}$$

Plugging both inequalities (12.4.9) and (12.4.10) into (12.4.8) we obtain a bound on the distance between parameters at time $t + 1$ as follows:

$$\|\mathbf{x}_t - \mathbf{x}^*\|^2 - \|\mathbf{x}_{t+1} - \mathbf{x}^*\|^2 \geq 2\eta_t (f(\boldsymbol{\xi}_t, \mathbf{x}_t) - f(\boldsymbol{\xi}_t, \mathbf{x}^*)) - \eta_t^2 L^2. \tag{12.4.11}$$

This means that we make progress as long as the difference between current loss and the optimal loss outweighs $\eta_t L^2/2$. Since this difference is bound to converge to zero it follows that the learning rate $\eta_t$ also needs to *vanish*.

Next we take expectations over (12.4.11). This yields

$$E\left[\|\mathbf{x}_t - \mathbf{x}^*\|^2\right] - E\left[\|\mathbf{x}_{t+1} - \mathbf{x}^*\|^2\right] \geq 2\eta_t [E[R(\mathbf{x}_t)] - R^*] - \eta_t^2 L^2. \tag{12.4.12}$$

The last step involves summing over the inequalities for $t \in \{1, \ldots, T\}$. Since the sum telescopes and by dropping the lower term we obtain

$$\|\mathbf{x}_1 - \mathbf{x}^*\|^2 \geq 2\left(\sum_{t=1}^{T} \eta_t\right)[E[R(\mathbf{x}_t)] - R^*] - L^2 \sum_{t=1}^{T} \eta_t^2. \tag{12.4.13}$$

Note that we exploited that $\mathbf{x}_1$ is given and thus the expectation can be dropped. Last define

$$\bar{\mathbf{x}} \stackrel{\mathrm{def}}{=} \frac{\sum_{t=1}^{T} \eta_t \mathbf{x}_t}{\sum_{t=1}^{T} \eta_t}. \tag{12.4.14}$$

Since

$$E\left(\frac{\sum_{t=1}^{T} \eta_t R(\mathbf{x}_t)}{\sum_{t=1}^{T} \eta_t}\right) = \frac{\sum_{t=1}^{T} \eta_t E[R(\mathbf{x}_t)]}{\sum_{t=1}^{T} \eta_t} = E[R(\mathbf{x}_t)], \tag{12.4.15}$$

by Jensen's inequality (setting $i = t$, $\alpha_i = \eta_t / \sum_{t=1}^{T} \eta_t$ in (12.2.3)) and convexity of $R$ it follows that $E[R(\mathbf{x}_t)] \geq E[R(\bar{\mathbf{x}})]$, thus

$$\sum_{t=1}^{T} \eta_t E[R(\mathbf{x}_t)] \geq \sum_{t=1}^{T} \eta_t E\left[R(\bar{\mathbf{x}})\right]. \tag{12.4.16}$$

Plugging this into the inequality (12.4.13) yields the bound

$$[E[\bar{\mathbf{x}}]] - R^* \leq \frac{r^2 + L^2 \sum_{t=1}^{T} \eta_t^2}{2 \sum_{t=1}^{T} \eta_t}, \tag{12.4.17}$$

where $r^2 \stackrel{\mathrm{def}}{=} \|\mathbf{x}_1 - \mathbf{x}^*\|^2$ is a bound on the distance between the initial choice of parameters and the final outcome. In short, the speed of convergence depends on how the norm of stochastic gradient is bounded ($L$) and how far away from optimality the initial parameter value is ($r$). Note that the bound is in terms of $\bar{\mathbf{x}}$ rather than $\mathbf{x}_T$. This is the case since $\bar{\mathbf{x}}$ is a smoothed version of the optimization path. Whenever $r$, $L$, and $T$ are known we can pick the learning rate $\eta = r/(L\sqrt{T})$. This yields as upper bound $rL/\sqrt{T}$. That is, we converge with rate $O(1/\sqrt{T})$ to the optimal solution.

### 12.4.4 Stochastic Gradients and Finite Samples

So far we have played a bit fast and loose when it comes to talking about stochastic gradient descent. We posited that we draw instances $x_i$, typically with labels $y_i$ from some distribution $p(x, y)$ and that we use this to update the model parameters in some manner. In particular, for a finite sample size we simply argued that the discrete distribution $p(x, y) = \frac{1}{n} \sum_{i=1}^n \delta_{x_i}(x)\delta_{y_i}(y)$ for some functions $\delta_{x_i}$ and $\delta_{y_i}$ allows us to perform stochastic gradient descent over it.

However, this is not really what we did. In the toy examples in the current section we simply added noise to an otherwise non-stochastic gradient, i.e., we pretended to have pairs $(x_i, y_i)$. It turns out that this is justified here (see the exercises for a detailed discussion). More troubling is that in all previous discussions we clearly did not do this. Instead we iterated over all instances *exactly once*. To see why this is preferable consider the converse, namely that we are sampling $n$ observations from the discrete distribution *with replacement*. The probability of choosing an element $i$ at random is $1/n$. Thus to choose it *at least* once is

$$P(\text{choose } i) = 1 - P(\text{omit } i) = 1 - (1 - 1/n)^n \approx 1 - e^{-1} \approx 0.63. \tag{12.4.18}$$

A similar reasoning shows that the probability of picking some sample (i.e., training example) *exactly once* is given by

$$\binom{n}{1} \frac{1}{n} \left(1 - \frac{1}{n}\right)^{n-1} = \frac{n}{n-1} \left(1 - \frac{1}{n}\right)^{n} \approx e^{-1} \approx 0.37. \tag{12.4.19}$$

Sampling with replacement leads to an increased variance and decreased data efficiency relative to sampling *without replacement*. Hence, in practice we perform the latter (and this is the default choice throughout this book). Last note that repeated passes through the training dataset traverse it in a *different* random order.

### 12.4.5 Summary

- For convex problems we can prove that for a wide choice of learning rates stochastic gradient descent will converge to the optimal solution.
- For deep learning this is generally not the case. However, the analysis of convex problems gives us useful insight into how to approach optimization, namely to reduce the learning rate progressively, albeit not too quickly.
- Problems occur when the learning rate is too small or too large. In practice a suitable learning rate is often found only after multiple experiments.
- When there are more examples in the training dataset, it costs more to compute each iteration for gradient descent, so stochastic gradient descent is preferred in these cases.
- Optimality guarantees for stochastic gradient descent are in general not available in nonconvex cases since the number of local minima that require checking might well be exponential.

### 12.4.6 Exercises

1. Experiment with different learning rate schedules for stochastic gradient descent and with different numbers of iterations. In particular, plot the distance from the optimal solution $(0, 0)$ as a function of the number of iterations.
2. Prove that for the function $f(x_1, x_2) = x_1^2 + 2x_2^2$ adding normal noise to the gradient is equivalent to minimizing a loss function $f(\mathbf{x}, \mathbf{w}) = (x_1 - w_1)^2 + 2(x_2 - w_2)^2$ where $\mathbf{x}$ is drawn from a normal distribution.
3. Compare convergence of stochastic gradient descent when you sample from $\{(x_1, y_1), \ldots, (x_n, y_n)\}$ with replacement and when you sample without replacement.
4. How would you change the stochastic gradient descent solver if some gradient (or rather some coordinate associated with it) was consistently larger than all the other gradients?
5. Assume that $f(x) = x^2(1 + \sin x)$. How many local minima does $f$ have? Can you change $f$ in such a way that to minimize it one needs to evaluate all the local minima?

Discussions[170].

## 12.5 Minibatch Stochastic Gradient Descent

So far we encountered two extremes in the approach to gradient-based learning: Section 12.3 uses the full dataset to compute gradients and to update parameters, one pass at a time. Conversely Section 12.4 processes one training example at a time to make progress. Either of them has its own drawbacks. Gradient descent is not particularly *data efficient* whenever data is very similar. Stochastic gradient descent is not particularly *computationally efficient* since CPUs and GPUs cannot exploit the full power of vectorization. This suggests that there might be something in between, and in fact, that is what we have been using so far in the examples we discussed.

### 12.5.1 Vectorization and Caches

At the heart of the decision to use minibatches is computational efficiency. This is most easily understood when considering parallelization to multiple GPUs and multiple servers. In this case we need to send at least one image to each GPU. With 8 GPUs per server and 16 servers we already arrive at a minibatch size no smaller than 128.

Things are a bit more subtle when it comes to single GPUs or even CPUs. These devices have multiple types of memory, often multiple types of computational units and different bandwidth constraints between them. For instance, a CPU has a small number of registers and then the L1, L2, and in some cases even L3 cache (which is shared among different processor cores). These caches are of increasing size and latency (and at the same time they are of decreasing bandwidth). Suffice to say, the processor is capable of performing many more operations than what the main memory interface is able to provide.

First, a 2GHz CPU with 16 cores and AVX-512 vectorization can process up to $2 \cdot 10^9 \cdot 16 \cdot 32 = 10^{12}$ bytes per second. The capability of GPUs easily exceeds this number by a factor of 100. On the other hand, a midrange server processor might not have much more than 100 GB/s bandwidth, i.e., less than one tenth of what would be required to keep the processor fed. To make matters worse, not all memory access is created equal: memory interfaces are typically 64 bit wide or wider (e.g., on GPUs up to 384 bit), hence reading a single byte incurs the cost of a much wider access.

Second, there is significant overhead for the first access whereas sequential access is relatively cheap (this is often called a burst read). There are many more things to keep in mind, such as caching when we have multiple sockets, chiplets, and other structures. See this Wikipedia article[171] for a more in-depth discussion.

The way to alleviate these constraints is to use a hierarchy of CPU caches that are actually fast enough to supply the processor with data. This is *the* driving force behind batching in deep learning. To keep matters simple, consider matrix-matrix multiplication, say $\mathbf{A} = \mathbf{B}\mathbf{C}$. We have a number of options for calculating $\mathbf{A}$. For instance, we could try the following:

1. We could compute $\mathbf{A}_{ij} = \mathbf{B}_{i,:}\mathbf{C}_{:,j}$, i.e., we could compute it elementwise by means of dot products.
2. We could compute $\mathbf{A}_{:,j} = \mathbf{B}\mathbf{C}_{:,j}$, i.e., we could compute it one column at a time. Likewise we could compute $\mathbf{A}$ one row $\mathbf{A}_{i,:}$ at a time.
3. We could simply compute $\mathbf{A} = \mathbf{B}\mathbf{C}$.
4. We could break $\mathbf{B}$ and $\mathbf{C}$ into smaller block matrices and compute $\mathbf{A}$ one block at a time.

If we follow the first option, we will need to copy one row and one column vector into the CPU each time we want to compute an element $\mathbf{A}_{ij}$. Even worse, due to the fact that matrix elements are aligned sequentially we are thus required to access many disjoint locations for one of the two vectors as we read them from memory. The second option is much more favorable. In it, we are able to keep the column vector $\mathbf{C}_{:,j}$ in the CPU cache while we keep on traversing through $\mathbf{B}$. This halves the memory bandwidth requirement with correspondingly faster access. Of course, option 3 is most desirable. Unfortunately, most matrices might not entirely fit into cache (this is what we are discussing after all). However, option 4 offers a practically useful alternative: we can move blocks of the matrix into cache and multiply them locally. Optimized libraries take care of this for us. Let's have a look at how efficient these operations are in practice.

Beyond computational efficiency, the overhead introduced by Python and by the deep learning framework itself is considerable. Recall that each time we execute a command the Python interpreter sends a command to the MXNet engine which needs to insert it into the computational graph and deal with it during scheduling. Such overhead can be quite detrimental. In short, it is highly advisable to use vectorization (and matrices) whenever possible.

```
%matplotlib inline
import time
import numpy as np
import torch
from torch import nn
from d2l import torch as d2l

A = torch.zeros(256, 256)
B = torch.randn(256, 256)
C = torch.randn(256, 256)
```

Since we will benchmark the running time frequently in the rest of the book, let's define a timer.

```
class Timer:  #@save
    """Record multiple running times."""
    def __init__(self):
        self.times = []
        self.start()

    def start(self):
        """Start the timer."""
        self.tik = time.time()

    def stop(self):
        """Stop the timer and record the time in a list."""
        self.times.append(time.time() - self.tik)
        return self.times[-1]

    def avg(self):
        """Return the average time."""
        return sum(self.times) / len(self.times)

    def sum(self):
        """Return the sum of time."""
        return sum(self.times)

    def cumsum(self):
        """Return the accumulated time."""
        return np.array(self.times).cumsum().tolist()

timer = Timer()
```

Element-wise assignment simply iterates over all rows and columns of $\mathbf{B}$ and $\mathbf{C}$ respectively to assign the value to $\mathbf{A}$.

```
# Compute A = BC one element at a time
timer.start()
for i in range(256):
    for j in range(256):
        A[i, j] = torch.dot(B[i, :], C[:, j])
timer.stop()
```

```
2.3525452613830566
```

A faster strategy is to perform column-wise assignment.

```
# Compute A = BC one column at a time
timer.start()
for j in range(256):
    A[:, j] = torch.mv(B, C[:, j])
timer.stop()
```

```
0.1891944408416748
```

Last, the most effective manner is to perform the entire operation in one block. Note that multiplying any two matrices $\mathbf{B} \in \mathbb{R}^{m \times n}$ and $\mathbf{C} \in \mathbb{R}^{n \times p}$ takes approximately $2mnp$ floating point operations, when scalar multiplication and addition are counted as separate operations (fused in practice). Thus, multiplying two $256 \times 256$ matrices takes $0.03$ billion floating point operations. Let's see what the respective speed of the operations is.

```
# Compute A = BC in one go
timer.start()
A = torch.mm(B, C)
timer.stop()

gigaflops = [0.03 / i for i in timer.times]
print(f'performance in Gigaflops: element {gigaflops[0]:.3f}, '
      f'column {gigaflops[1]:.3f}, full {gigaflops[2]:.3f}')
```

```
performance in Gigaflops: element 0.013, column 0.159, full 3.200
```

## 12.5.2 Minibatches

In the past we took it for granted that we would read *minibatches* of data rather than single observations to update parameters. We now give a brief justification for it. Processing single observations requires us to perform many single matrix-vector (or even vector-vector) multiplications, which is quite expensive and which incurs a significant overhead on behalf of the underlying deep learning framework. This applies both to evaluating a network when applied to data (often referred to as inference) and when computing gradients to update parameters. That is, this applies whenever we perform $\mathbf{w} \leftarrow \mathbf{w} - \eta_t \mathbf{g}_t$ where

$$\mathbf{g}_t = \partial_{\mathbf{w}} f(\mathbf{x}_t, \mathbf{w}) \tag{12.5.1}$$

We can increase the *computational* efficiency of this operation by applying it to a minibatch of observations at a time. That is, we replace the gradient $\mathbf{g}_t$ over a single observation by one

over a small batch

$$\mathbf{g}_t = \partial_{\mathbf{w}} \frac{1}{|\mathcal{B}_t|} \sum_{i \in \mathcal{B}_t} f(\mathbf{x}_i, \mathbf{w}) \tag{12.5.2}$$

Let's see what this does to the statistical properties of $\mathbf{g}_t$: since both $\mathbf{x}_t$ and also all elements of the minibatch $\mathcal{B}_t$ are drawn uniformly at random from the training set, the expectation of the gradient remains unchanged. The variance, on the other hand, is reduced significantly. Since the minibatch gradient is composed of $b \stackrel{\textrm{def}}{=} |\mathcal{B}_t|$ independent gradients which are being averaged, its standard deviation is reduced by a factor of $b^{-\frac{1}{2}}$. This, by itself, is a good thing, since it means that the updates are more reliably aligned with the full gradient.

Naively this would indicate that choosing a large minibatch $\mathcal{B}_t$ would be universally desirable. Alas, after some point, the additional reduction in standard deviation is minimal when compared to the linear increase in computational cost. In practice we pick a minibatch that is large enough to offer good computational efficiency while still fitting into the memory of a GPU. To illustrate the savings let's have a look at some code. In it we perform the same matrix-matrix multiplication, but this time broken up into "minibatches" of 64 columns at a time.

```
timer.start()
for j in range(0, 256, 64):
    A[:, j:j+64] = torch.mm(B, C[:, j:j+64])
timer.stop()
print(f'performance in Gigaflops: block {0.03 / timer.times[3]:.3f}')
```

```
performance in Gigaflops: block 1.258
```

As we can see, the computation on the minibatch is essentially as efficient as on the full matrix. A word of caution is in order. In Section 8.5 we used a type of regularization that was heavily dependent on the amount of variance in a minibatch. As we increase the latter, the variance decreases and with it the benefit of the noise-injection due to batch normalization. See e.g., Ioffe (2017) for details on how to rescale and compute the appropriate terms.

## 12.5.3 Reading the Dataset

Let's have a look at how minibatches are efficiently generated from data. In the following we use a dataset developed by NASA to test the wing noise from different aircraft[172] to compare these optimization algorithms. For convenience we only use the first $1,500$ examples. The data is whitened for preprocessing, i.e., we remove the mean and rescale the variance to $1$ per coordinate.

```
#@save
d2l.DATA_HUB['airfoil'] = (d2l.DATA_URL + 'airfoil_self_noise.dat',
                           '76e5be1548fd8222e5074cf0faae75edff8cf93f')
```

```
#@save
def get_data_ch11(batch_size=10, n=1500):
    data = np.genfromtxt(d2l.download('airfoil'),
                         dtype=np.float32, delimiter='\t')
    data = torch.from_numpy((data - data.mean(axis=0)) / data.std(axis=0))
    data_iter = d2l.load_array((data[:n, :-1], data[:n, -1]),
                               batch_size, is_train=True)
    return data_iter, data.shape[1]-1
```

## 12.5.4 Implementation from Scratch

Recall the minibatch stochastic gradient descent implementation from Section 3.4. In the following we provide a slightly more general implementation. For convenience it has the same call signature as the other optimization algorithms introduced later in this chapter. Specifically, we add the status input `states` and place the hyperparameter in dictionary `hyperparams`. In addition, we will average the loss of each minibatch example in the training function, so the gradient in the optimization algorithm does not need to be divided by the batch size.

```
def sgd(params, states, hyperparams):
    for p in params:
        p.data.sub_(hyperparams['lr'] * p.grad)
        p.grad.data.zero_()
```

Next, we implement a generic training function to facilitate the use of the other optimization algorithms introduced later in this chapter. It initializes a linear regression model and can be used to train the model with minibatch stochastic gradient descent and other algorithms introduced subsequently.

```
#@save
def train_ch11(trainer_fn, states, hyperparams, data_iter,
               feature_dim, num_epochs=2):
    # Initialization
    w = torch.normal(mean=0.0, std=0.01, size=(feature_dim, 1),
                     requires_grad=True)
    b = torch.zeros((1), requires_grad=True)
    net, loss = lambda X: d2l.linreg(X, w, b), d2l.squared_loss
    # Train
    animator = d2l.Animator(xlabel='epoch', ylabel='loss',
                            xlim=[0, num_epochs], ylim=[0.22, 0.35])
    n, timer = 0, d2l.Timer()
    for _ in range(num_epochs):
        for X, y in data_iter:
            l = loss(net(X), y).mean()
            l.backward()
            trainer_fn([w, b], states, hyperparams)
            n += X.shape[0]
            if n % 200 == 0:
```

```
                timer.stop()
                animator.add(n/X.shape[0]/len(data_iter),
                             (d2l.evaluate_loss(net, data_iter, loss),))
                timer.start()
    print(f'loss: {animator.Y[0][-1]:.3f}, {timer.sum()/num_epochs:.3f} sec/
↪epoch')
    return timer.cumsum(), animator.Y[0]
```

Let's see how optimization proceeds for batch gradient descent. This can be achieved by setting the minibatch size to 1500 (i.e., to the total number of examples). As a result the model parameters are updated only once per epoch. There is little progress. In fact, after 6 steps progress stalls.

```
def train_sgd(lr, batch_size, num_epochs=2):
    data_iter, feature_dim = get_data_ch11(batch_size)
    return train_ch11(
        sgd, None, {'lr': lr}, data_iter, feature_dim, num_epochs)

gd_res = train_sgd(1, 1500, 10)
```

```
loss: 0.253, 0.028 sec/epoch
```

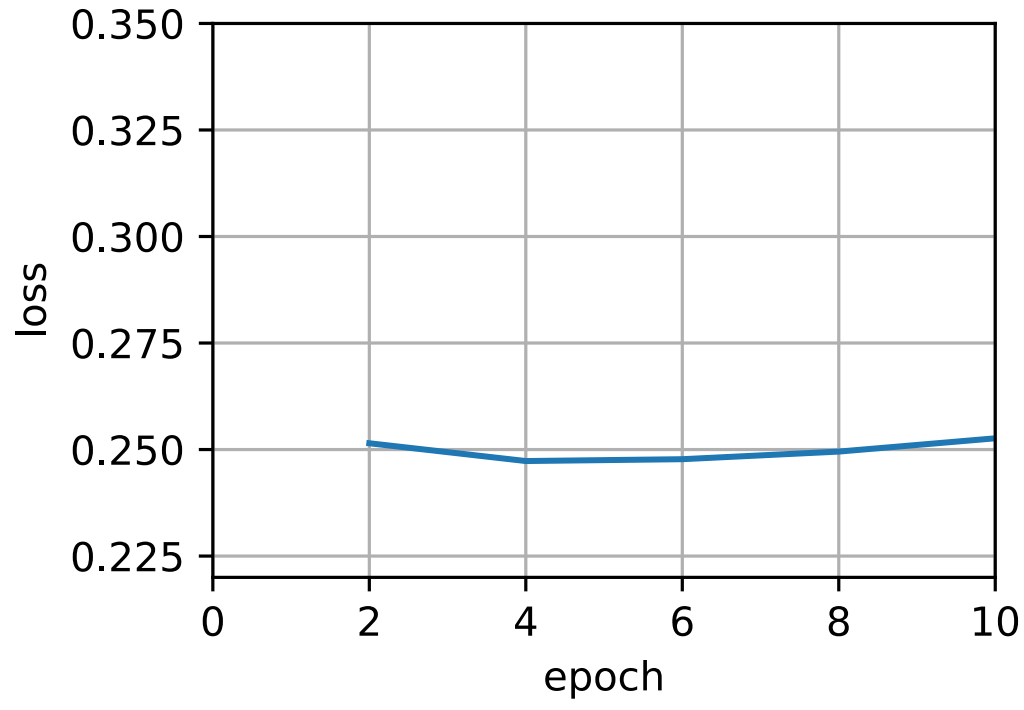

When the batch size equals 1, we use stochastic gradient descent for optimization. For simplicity of implementation we picked a constant (albeit small) learning rate. In stochastic gradient descent, the model parameters are updated whenever an example is processed. In our case this amounts to 1500 updates per epoch. As we can see, the decline in the value of the objective function slows down after one epoch. Although both the procedures processed 1500 examples within one epoch, stochastic gradient descent consumes more time than gradient descent in our experiment. This is because stochastic gradient descent updated the parameters more frequently and since it is less efficient to process single observations one at a time.

```
sgd_res = train_sgd(0.005, 1)
```

```
loss: 0.243, 0.954 sec/epoch
```

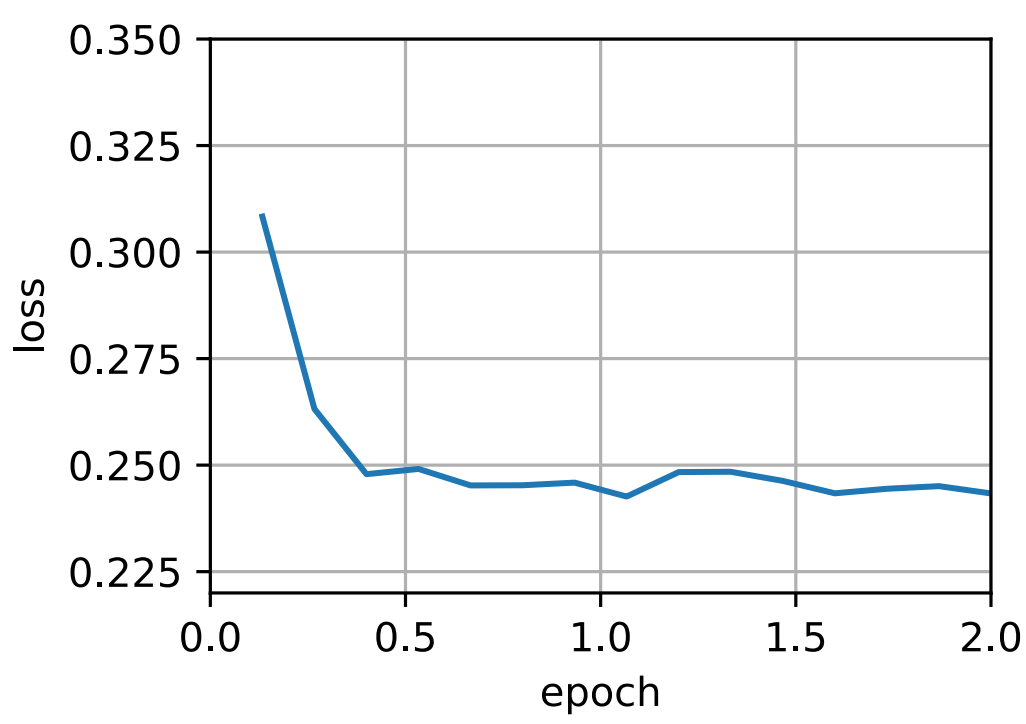

Finally, when the batch size equals 100, we use minibatch stochastic gradient descent for optimization. The time required per epoch is shorter than the time needed for stochastic gradient descent and the time for batch gradient descent.

```
mini1_res = train_sgd(.4, 100)
```

```
loss: 0.246, 0.035 sec/epoch
```

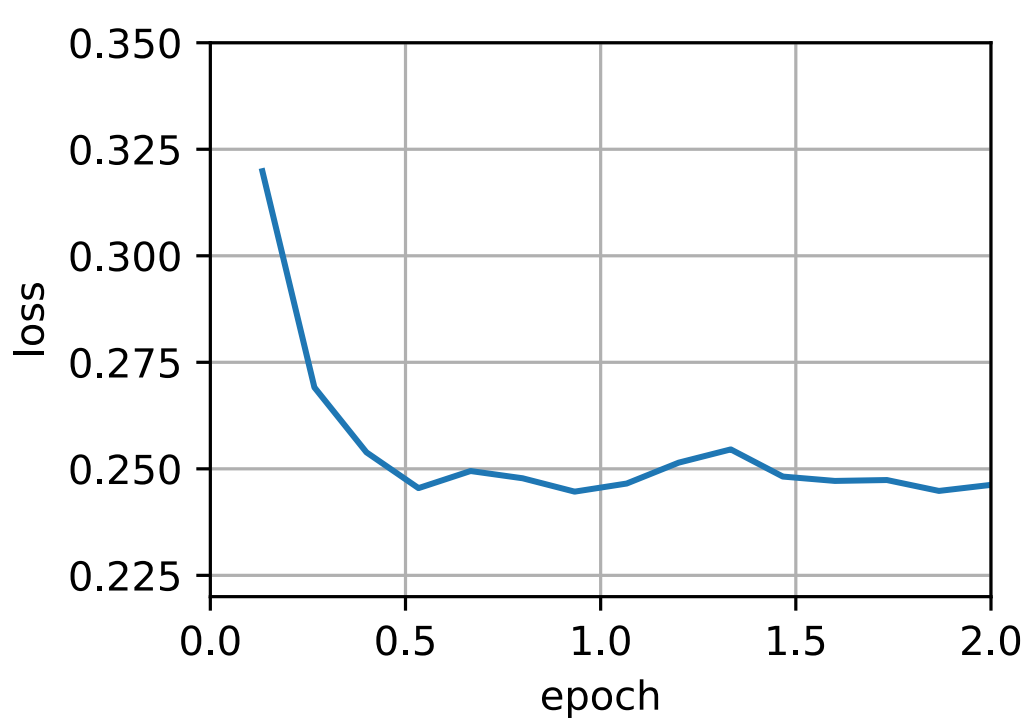

Reducing the batch size to 10, the time for each epoch increases because the workload for each batch is less efficient to execute.

```
mini2_res = train_sgd(.05, 10)
```

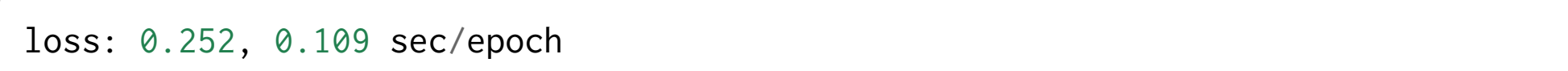

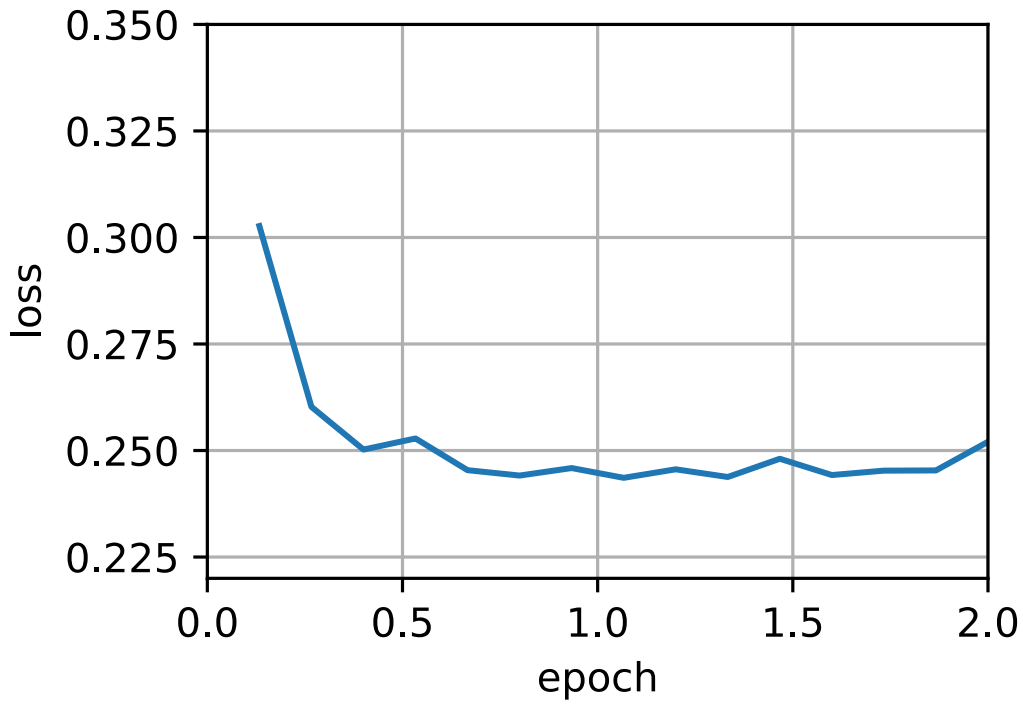

Now we can compare the time vs. loss for the previous four experiments. As can be seen, although stochastic gradient descent converges faster than GD in terms of number of examples processed, it uses more time to reach the same loss than GD because computing the gradient example by example is not as efficient. Minibatch stochastic gradient descent is able to trade-off convergence speed and computation efficiency. A minibatch size of 10 is more efficient than stochastic gradient descent; a minibatch size of 100 even outperforms GD in terms of runtime.

```
d2l.set_figsize([6, 3])
d2l.plot(*list(map(list, zip(gd_res, sgd_res, mini1_res, mini2_res))),
         'time (sec)', 'loss', xlim=[1e-2, 10],
         legend=['gd', 'sgd', 'batch size=100', 'batch size=10'])
d2l.plt.gca().set_xscale('log')
```

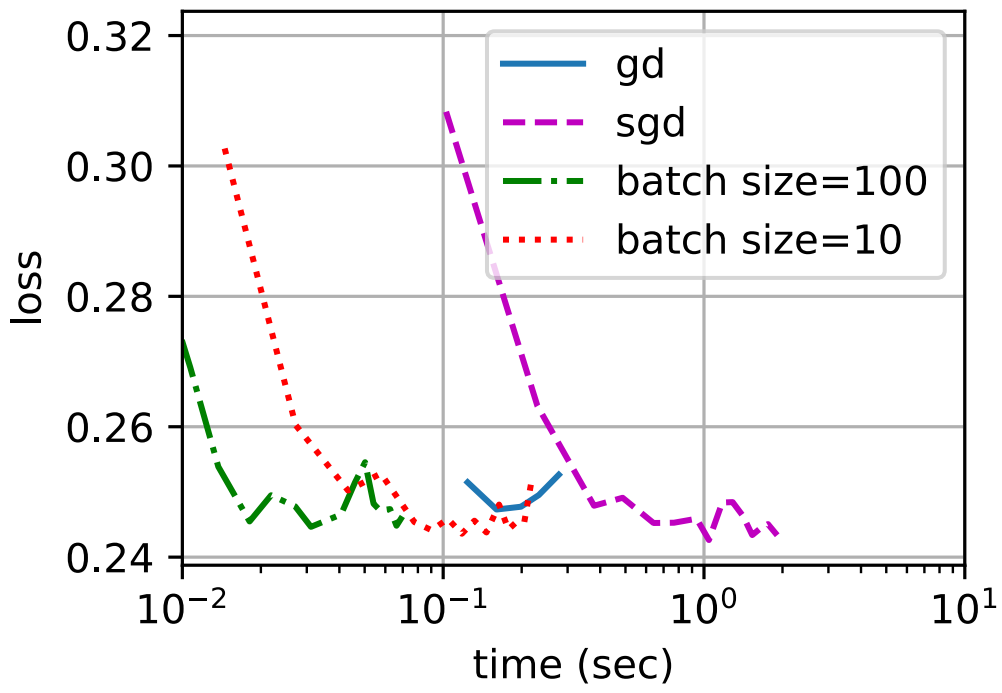

## 12.5.5 Concise Implementation

In Gluon, we can use the `Trainer` class to call optimization algorithms. This is used to implement a generic training function. We will use this throughout the current chapter.

```
#@save
def train_concise_ch11(trainer_fn, hyperparams, data_iter, num_epochs=4):
    # Initialization
    net = nn.Sequential(nn.Linear(5, 1))
    def init_weights(module):
        if type(module) == nn.Linear:
            torch.nn.init.normal_(module.weight, std=0.01)
    net.apply(init_weights)

    optimizer = trainer_fn(net.parameters(), **hyperparams)
    loss = nn.MSELoss(reduction='none')
    animator = d2l.Animator(xlabel='epoch', ylabel='loss',
                            xlim=[0, num_epochs], ylim=[0.22, 0.35])
    n, timer = 0, d2l.Timer()
    for _ in range(num_epochs):
        for X, y in data_iter:
            optimizer.zero_grad()
            out = net(X)
            y = y.reshape(out.shape)
            l = loss(out, y)
            l.mean().backward()
            optimizer.step()
            n += X.shape[0]
            if n % 200 == 0:
                timer.stop()
                # `MSELoss` computes squared error without the 1/2 factor
                animator.add(n/X.shape[0]/len(data_iter),
                             (d2l.evaluate_loss(net, data_iter, loss) / 2,))
                timer.start()
    print(f'loss: {animator.Y[0][-1]:.3f}, {timer.sum()/num_epochs:.3f} sec/
↪epoch')
```

Using Gluon to repeat the last experiment shows identical behavior.

```
data_iter, _ = get_data_ch11(10)
trainer = torch.optim.SGD
train_concise_ch11(trainer, {'lr': 0.01}, data_iter)
```

```
loss: 0.242, 0.128 sec/epoch
```

### 12.5.6 Summary

- Vectorization makes code more efficient due to reduced overhead arising from the deep learning framework and due to better memory locality and caching on CPUs and GPUs.
- There is a trade-off between statistical efficiency arising from stochastic gradient descent and computational efficiency arising from processing large batches of data at a time.
- Minibatch stochastic gradient descent offers the best of both worlds: computational and statistical efficiency.

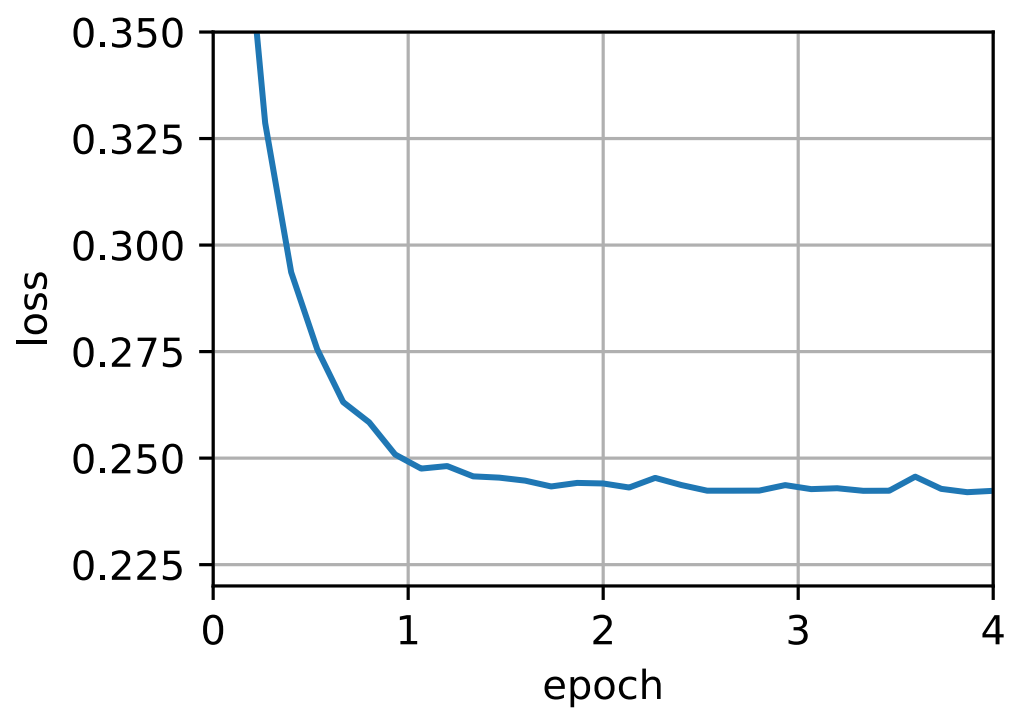

- In minibatch stochastic gradient descent we process batches of data obtained by a random permutation of the training data (i.e., each observation is processed only once per epoch, albeit in random order).
- It is advisable to decay the learning rates during training.
- In general, minibatch stochastic gradient descent is faster than stochastic gradient descent and gradient descent for convergence to a smaller risk, when measured in terms of clock time.

### 12.5.7 Exercises

1. Modify the batch size and learning rate and observe the rate of decline for the value of the objective function and the time consumed in each epoch.
2. Read the MXNet documentation and use the `Trainer` class `set_learning_rate` function to reduce the learning rate of the minibatch stochastic gradient descent to 1/10 of its previous value after each epoch.
3. Compare minibatch stochastic gradient descent with a variant that actually *samples with replacement* from the training set. What happens?
4. An evil genie replicates your dataset without telling you (i.e., each observation occurs twice and your dataset grows to twice its original size, but nobody told you). How does the behavior of stochastic gradient descent, minibatch stochastic gradient descent and that of gradient descent change?

Discussions[173].

# 12.6 Momentum

In Section 12.4 we reviewed what happens when performing stochastic gradient descent, i.e., when performing optimization where only a noisy variant of the gradient is available. In particular, we noticed that for noisy gradients we need to be extra cautious when it comes to choosing the learning rate in the face of noise. If we decrease it too rapidly, convergence stalls. If we are too lenient, we fail to converge to a good enough solution since noise keeps on driving us away from optimality.

## 12.6.1 Basics

In this section, we will explore more effective optimization algorithms, especially for certain types of optimization problems that are common in practice.

### Leaky Averages

The previous section saw us discussing minibatch SGD as a means for accelerating computation. It also had the nice side-effect that averaging gradients reduced the amount of variance. The minibatch stochastic gradient descent can be calculated by:

$$\mathbf{g}_{t,t-1} = \partial_{\mathbf{w}} \frac{1}{|\mathcal{B}_t|} \sum_{i \in \mathcal{B}_t} f(\mathbf{x}_i, \mathbf{w}_{t-1}) = \frac{1}{|\mathcal{B}_t|} \sum_{i \in \mathcal{B}_t} \mathbf{h}_{i,t-1}. \tag{12.6.1}$$

To keep the notation simple, here we used $\mathbf{h}_{i,t-1} = \partial_{\mathbf{w}} f(\mathbf{x}_i, \mathbf{w}_{t-1})$ as the stochastic gradient descent for sample $i$ using the weights updated at time $t-1$. It would be nice if we could benefit from the effect of variance reduction even beyond averaging gradients on a minibatch. One option to accomplish this task is to replace the gradient computation by a "leaky average":

$$\mathbf{v}_t = \beta \mathbf{v}_{t-1} + \mathbf{g}_{t,t-1} \tag{12.6.2}$$

for some $\beta \in (0, 1)$. This effectively replaces the instantaneous gradient by one that is been averaged over multiple *past* gradients. $\mathbf{v}$ is called *velocity*. It accumulates past gradients similar to how a heavy ball rolling down the objective function landscape integrates over past forces. To see what is happening in more detail let's expand $\mathbf{v}_t$ recursively into

$$\mathbf{v}_t = \beta^2 \mathbf{v}_{t-2} + \beta \mathbf{g}_{t-1,t-2} + \mathbf{g}_{t,t-1} = \ldots, = \sum_{\tau=0}^{t-1} \beta^\tau \mathbf{g}_{t-\tau,t-\tau-1}. \tag{12.6.3}$$

Large $\beta$ amounts to a long-range average, whereas small $\beta$ amounts to only a slight correction relative to a gradient method. The new gradient replacement no longer points into the direction of steepest descent on a particular instance any longer but rather in the direction of a weighted average of past gradients. This allows us to realize most of the benefits of averaging over a batch without the cost of actually computing the gradients on it. We will revisit this averaging procedure in more detail later.

The above reasoning formed the basis for what is now known as *accelerated* gradient methods, such as gradients with momentum. They enjoy the additional benefit of being much more effective in cases where the optimization problem is ill-conditioned (i.e., where there are some directions where progress is much slower than in others, resembling a narrow canyon). Furthermore, they allow us to average over subsequent gradients to obtain more stable directions of descent. Indeed, the aspect of acceleration even for noise-free convex problems is one of the key reasons why momentum works and why it works so well.

As one would expect, due to its efficacy momentum is a well-studied subject in optimization for deep learning and beyond. See e.g., the beautiful expository article[174] by Goh (2017) for an in-depth analysis and interactive animation. It was proposed by Polyak (1964). Nesterov (2018) has a detailed theoretical discussion in the context of convex optimization. Momentum in deep learning has been known to be beneficial for a long time. See e.g., the discussion by Sutskever *et al.* (2013) for details.

### An Ill-conditioned Problem

To get a better understanding of the geometric properties of the momentum method we revisit gradient descent, albeit with a significantly less pleasant objective function. Recall that in Section 12.3 we used $f(\mathbf{x}) = x_1^2 + 2x_2^2$, i.e., a moderately distorted ellipsoid objective. We distort this function further by stretching it out in the $x_1$ direction via

$$f(\mathbf{x}) = 0.1x_1^2 + 2x_2^2. \tag{12.6.4}$$

As before $f$ has its minimum at $(0, 0)$. This function is *very* flat in the direction of $x_1$. Let's see what happens when we perform gradient descent as before on this new function. We pick a learning rate of $0.4$.

```
%matplotlib inline
import torch
from d2l import torch as d2l

eta = 0.4
def f_2d(x1, x2):
    return 0.1 * x1 ** 2 + 2 * x2 ** 2
def gd_2d(x1, x2, s1, s2):
    return (x1 - eta * 0.2 * x1, x2 - eta * 4 * x2, 0, 0)

d2l.show_trace_2d(f_2d, d2l.train_2d(gd_2d))
```

```
epoch 20, x1: -0.943467, x2: -0.000073
```

By construction, the gradient in the $x_2$ direction is *much* higher and changes much more rapidly than in the horizontal $x_1$ direction. Thus we are stuck between two undesirable choices:

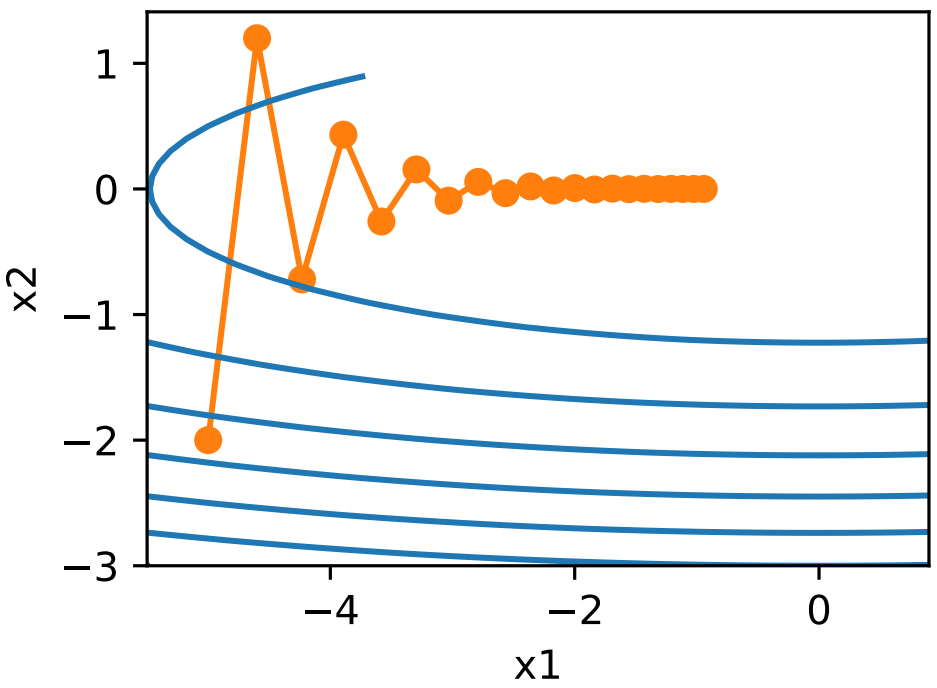

if we pick a small learning rate we ensure that the solution does not diverge in the $x_2$ direction but we are saddled with slow convergence in the $x_1$ direction. Conversely, with a large learning rate we progress rapidly in the $x_1$ direction but diverge in $x_2$. The example below illustrates what happens even after a slight increase in learning rate from $0.4$ to $0.6$. Convergence in the $x_1$ direction improves but the overall solution quality is much worse.

```
eta = 0.6
d2l.show_trace_2d(f_2d, d2l.train_2d(gd_2d))
```

```
epoch 20, x1: -0.387814, x2: -1673.365109
```

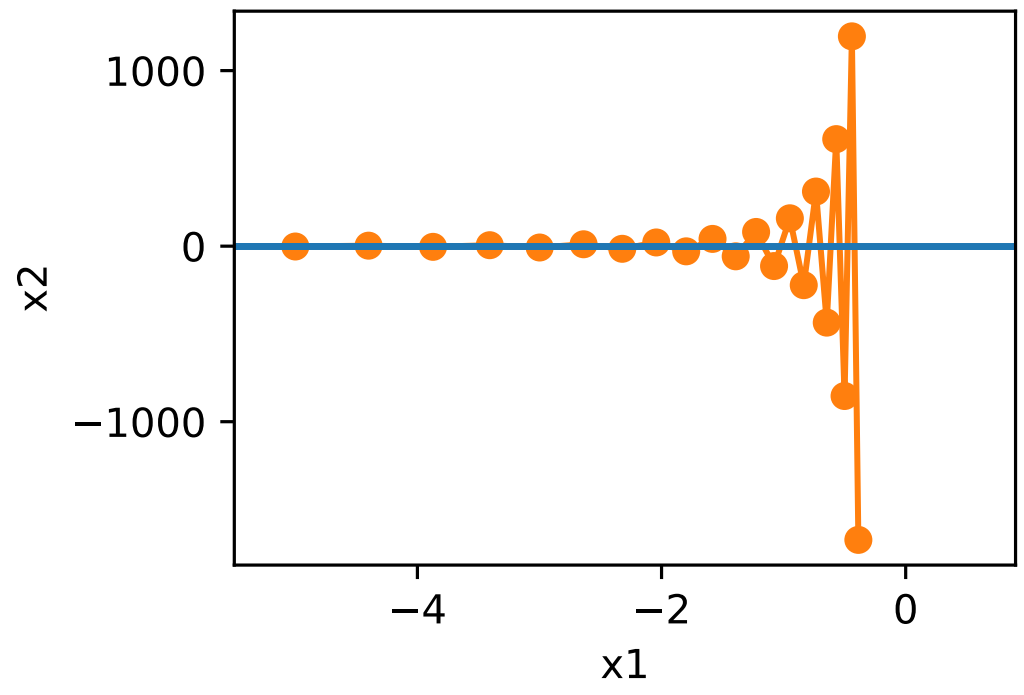

### The Momentum Method

The momentum method allows us to solve the gradient descent problem described above. Looking at the optimization trace above we might intuit that averaging gradients over the past would work well. After all, in the $x_1$ direction this will aggregate well-aligned gradients, thus increasing the distance we cover with every step. Conversely, in the $x_2$ direction where gradients oscillate, an aggregate gradient will reduce step size due to oscillations that cancel each other out. Using $\mathbf{v}_t$ instead of the gradient $\mathbf{g}_t$ yields the following update equations:

$$\begin{aligned}\mathbf{v}_t &\leftarrow \beta \mathbf{v}_{t-1} + \mathbf{g}_{t, t-1}, \\ \mathbf{x}_t &\leftarrow \mathbf{x}_{t-1} - \eta_t \mathbf{v}_t.\end{aligned} \tag{12.6.5}$$

Note that for $\beta = 0$ we recover regular gradient descent. Before delving deeper into the mathematical properties let's have a quick look at how the algorithm behaves in practice.

```
def momentum_2d(x1, x2, v1, v2):
    v1 = beta * v1 + 0.2 * x1
    v2 = beta * v2 + 4 * x2
    return x1 - eta * v1, x2 - eta * v2, v1, v2

eta, beta = 0.6, 0.5
d2l.show_trace_2d(f_2d, d2l.train_2d(momentum_2d))
```

```
epoch 20, x1: 0.007188, x2: 0.002553
```

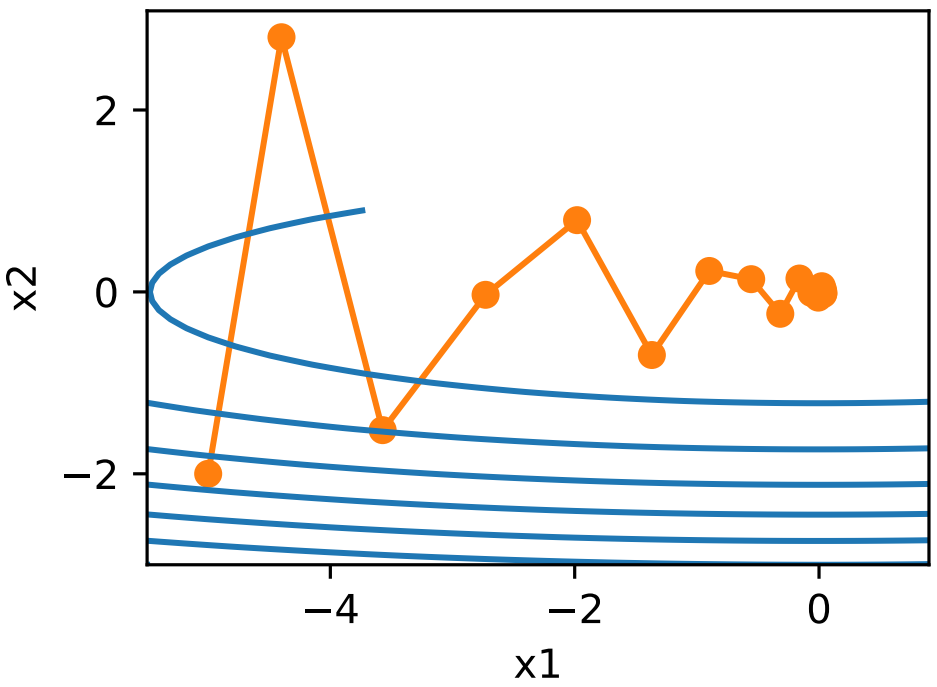

As we can see, even with the same learning rate that we used before, momentum still converges well. Let's see what happens when we decrease the momentum parameter. Halving it to $\beta = 0.25$ leads to a trajectory that barely converges at all. Nonetheless, it is a lot better than without momentum (when the solution diverges).

```
eta, beta = 0.6, 0.25
d2l.show_trace_2d(f_2d, d2l.train_2d(momentum_2d))
```

```
epoch 20, x1: -0.126340, x2: -0.186632
```

Note that we can combine momentum with stochastic gradient descent and in particular, minibatch stochastic gradient descent. The only change is that in that case we replace the gradients $\mathbf{g}_{t,t-1}$ with $\mathbf{g}_t$. Last, for convenience we initialize $\mathbf{v}_0 = 0$ at time $t = 0$. Let's look at what leaky averaging actually does to the updates.

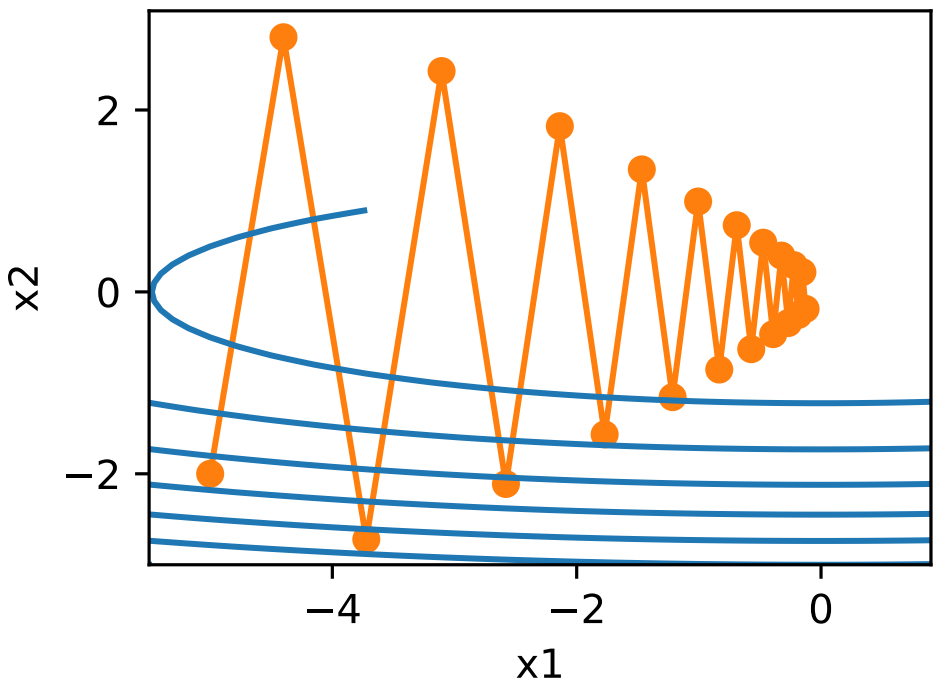

### Effective Sample Weight

Recall that $\mathbf{v}_t = \sum_{\tau=0}^{t-1} \beta^\tau \mathbf{g}_{t-\tau,t-\tau-1}$. In the limit the terms add up to $\sum_{\tau=0}^{\infty} \beta^\tau = \frac{1}{1-\beta}$. In other words, rather than taking a step of size $\eta$ in gradient descent or stochastic gradient descent we take a step of size $\frac{\eta}{1-\beta}$ while at the same time, dealing with a potentially much better behaved descent direction. These are two benefits in one. To illustrate how weighting behaves for different choices of $\beta$ consider the diagram below.

```
d2l.set_figsize()
betas = [0.95, 0.9, 0.6, 0]
for beta in betas:
    x = torch.arange(40).detach().numpy()
    d2l.plt.plot(x, beta ** x, label=f'beta = {beta:.2f}')
d2l.plt.xlabel('time')
d2l.plt.legend();
```

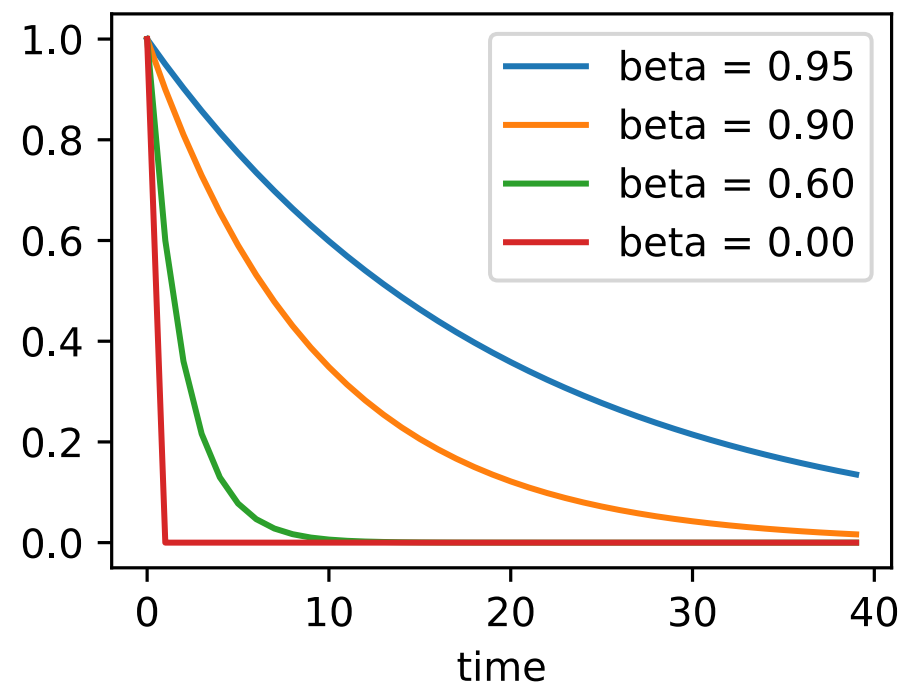

## 12.6.2 Practical Experiments

Let's see how momentum works in practice, i.e., when used within the context of a proper optimizer. For this we need a somewhat more scalable implementation.

### Implementation from Scratch

Compared with (minibatch) stochastic gradient descent the momentum method needs to maintain a set of auxiliary variables, i.e., velocity. It has the same shape as the gradients (and variables of the optimization problem). In the implementation below we call these variables `states`.

```
def init_momentum_states(feature_dim):
    v_w = torch.zeros((feature_dim, 1))
    v_b = torch.zeros(1)
    return (v_w, v_b)
```

```
def sgd_momentum(params, states, hyperparams):
    for p, v in zip(params, states):
        with torch.no_grad():
            v[:] = hyperparams['momentum'] * v + p.grad
            p[:] -= hyperparams['lr'] * v
        p.grad.data.zero_()
```

Let's see how this works in practice.

```
def train_momentum(lr, momentum, num_epochs=2):
    d2l.train_ch11(sgd_momentum, init_momentum_states(feature_dim),
                   {'lr': lr, 'momentum': momentum}, data_iter,
                   feature_dim, num_epochs)

data_iter, feature_dim = d2l.get_data_ch11(batch_size=10)
train_momentum(0.02, 0.5)
```

```
loss: 0.246, 0.195 sec/epoch
```

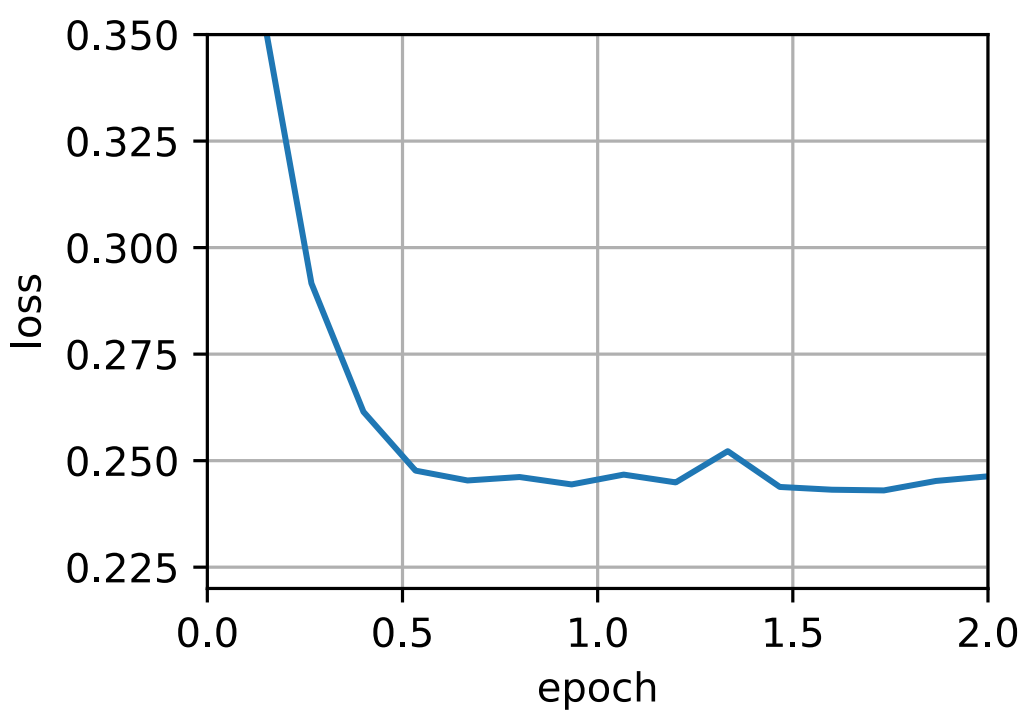

When we increase the momentum hyperparameter `momentum` to 0.9, it amounts to a significantly larger effective sample size of $\frac{1}{1-0.9} = 10$. We reduce the learning rate slightly to $0.01$ to keep matters under control.

```
train_momentum(0.01, 0.9)
```

```
loss: 0.254, 0.146 sec/epoch
```

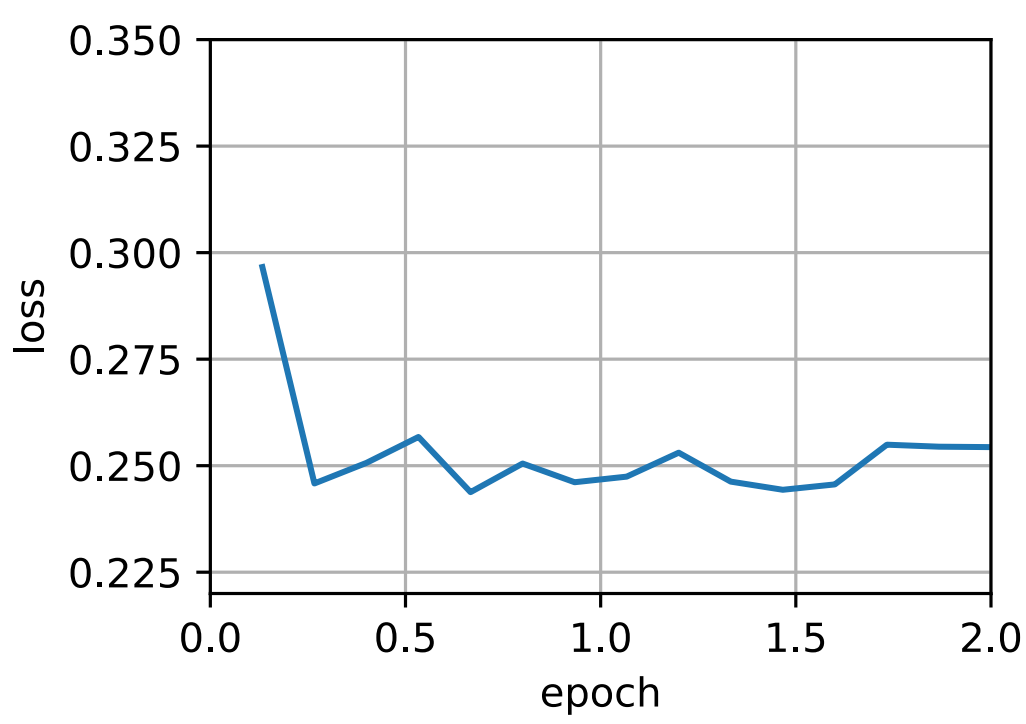

Reducing the learning rate further addresses any issue of non-smooth optimization problems. Setting it to $0.005$ yields good convergence properties.

```
train_momentum(0.005, 0.9)
```

```
loss: 0.247, 0.152 sec/epoch
```

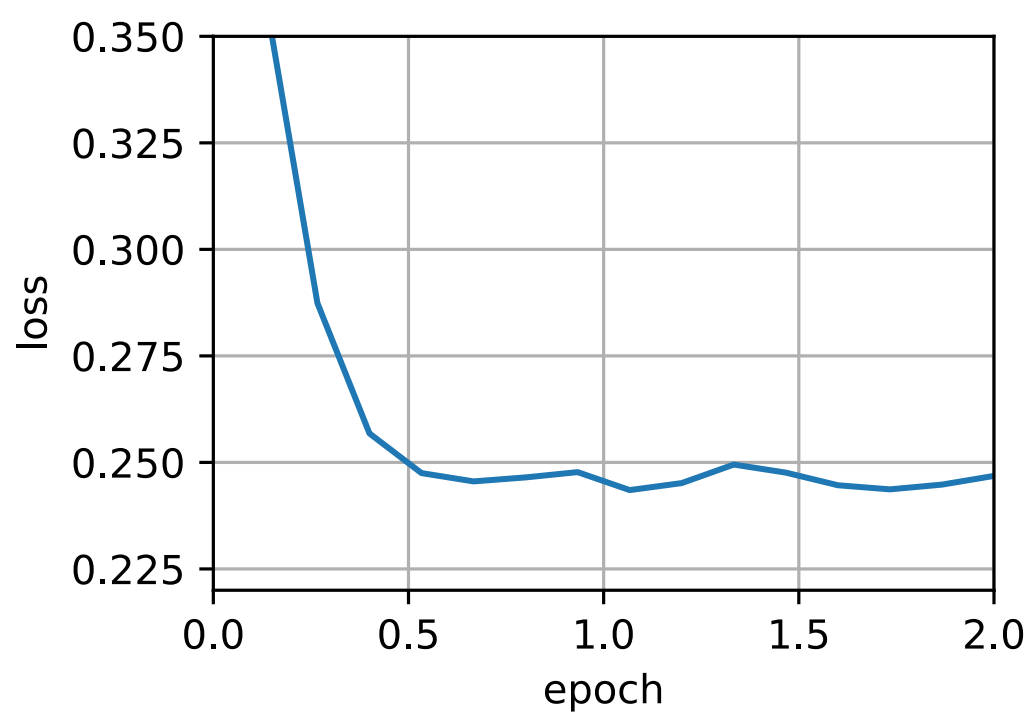

### Concise Implementation

There is very little to do in Gluon since the standard sgd solver already had momentum built in. Setting matching parameters yields a very similar trajectory.

```
trainer = torch.optim.SGD
d2l.train_concise_ch11(trainer, {'lr': 0.005, 'momentum': 0.9}, data_iter)
```

```
loss: 0.247, 0.142 sec/epoch
```

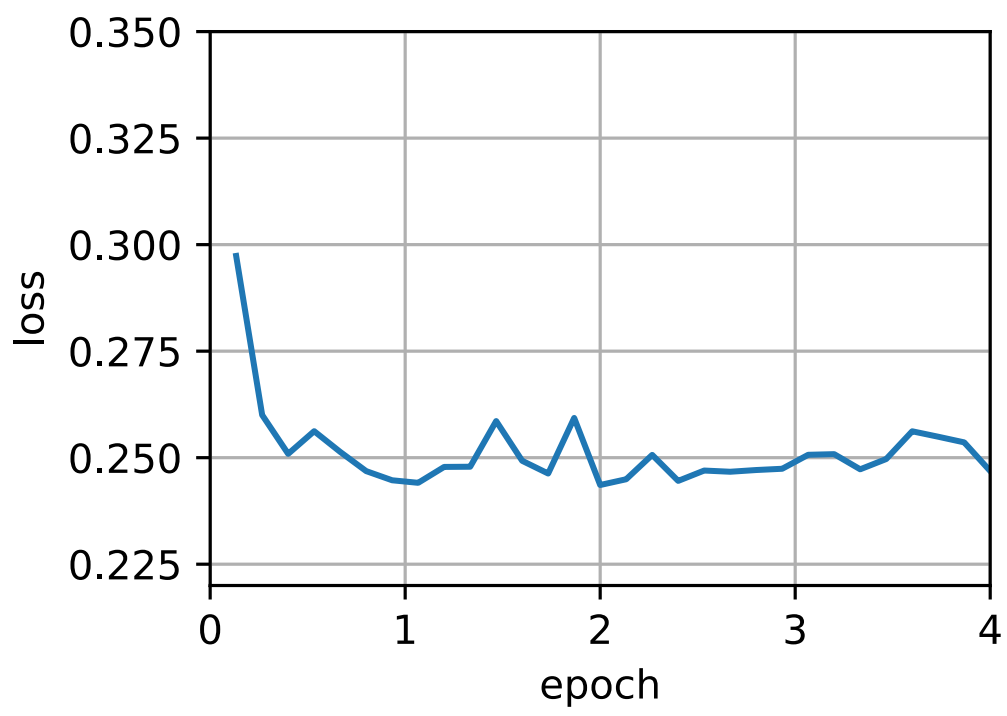

## 12.6.3 Theoretical Analysis

So far the 2D example of $f(x) = 0.1x_1^2 + 2x_2^2$ seemed rather contrived. We will now see that this is actually quite representative of the types of problem one might encounter, at least in the case of minimizing convex quadratic objective functions.

### Quadratic Convex Functions

Consider the function

$$h(\mathbf{x}) = \frac{1}{2}\mathbf{x}^\top \mathbf{Q}\mathbf{x} + \mathbf{x}^\top \mathbf{c} + b. \tag{12.6.6}$$

This is a general quadratic function. For positive definite matrices $\mathbf{Q} \succ 0$, i.e., for matrices with positive eigenvalues this has a minimizer at $\mathbf{x}^* = -\mathbf{Q}^{-1}\mathbf{c}$ with minimum value $b - \frac{1}{2}\mathbf{c}^\top \mathbf{Q}^{-1}\mathbf{c}$. Hence we can rewrite $h$ as

$$h(\mathbf{x}) = \frac{1}{2}(\mathbf{x} - \mathbf{Q}^{-1}\mathbf{c})^\top \mathbf{Q}(\mathbf{x} - \mathbf{Q}^{-1}\mathbf{c}) + b - \frac{1}{2}\mathbf{c}^\top \mathbf{Q}^{-1}\mathbf{c}. \tag{12.6.7}$$

The gradient is given by $\partial_{\mathbf{x}} h(\mathbf{x}) = \mathbf{Q}(\mathbf{x} - \mathbf{Q}^{-1}\mathbf{c})$. That is, it is given by the distance between $\mathbf{x}$ and the minimizer, multiplied by $\mathbf{Q}$. Consequently also the velocity is a linear combination of terms $\mathbf{Q}(\mathbf{x}_t - \mathbf{Q}^{-1}\mathbf{c})$.

Since $\mathbf{Q}$ is positive definite it can be decomposed into its eigensystem via $\mathbf{Q} = \mathbf{O}^\top \boldsymbol{\Lambda}\mathbf{O}$ for an orthogonal (rotation) matrix $\mathbf{O}$ and a diagonal matrix $\boldsymbol{\Lambda}$ of positive eigenvalues. This allows us to perform a change of variables from $\mathbf{x}$ to $\mathbf{z} \stackrel{\text{def}}{=} \mathbf{O}(\mathbf{x}-\mathbf{Q}^{-1}\mathbf{c})$ to obtain a much simplified expression:

$$h(\mathbf{z}) = \frac{1}{2}\mathbf{z}^\top \boldsymbol{\Lambda} \mathbf{z} + b'. \tag{12.6.8}$$

Here $b' = b - \frac{1}{2}\mathbf{c}^\top \mathbf{Q}^{-1}\mathbf{c}$. Since $\mathbf{O}$ is only an orthogonal matrix this does not perturb the gradients in a meaningful way. Expressed in terms of $\mathbf{z}$ gradient descent becomes

$$\mathbf{z}_t = \mathbf{z}_{t-1} - \boldsymbol{\Lambda} \mathbf{z}_{t-1} = (\mathbf{I} - \boldsymbol{\Lambda}) \mathbf{z}_{t-1}. \tag{12.6.9}$$

The important fact in this expression is that gradient descent *does not mix* between different eigenspaces. That is, when expressed in terms of the eigensystem of $\mathbf{Q}$ the optimization problem proceeds in a coordinate-wise manner. This also holds for

$$\begin{aligned}
\mathbf{v}_t & = \beta \mathbf{v}_{t-1} + \boldsymbol{\Lambda} \mathbf{z}_{t-1} \\
\mathbf{z}_t & = \mathbf{z}_{t-1} - \eta \left(\beta \mathbf{v}_{t-1} + \boldsymbol{\Lambda} \mathbf{z}_{t-1}\right) \\
& = (\mathbf{I} - \eta \boldsymbol{\Lambda}) \mathbf{z}_{t-1} - \eta \beta \mathbf{v}_{t-1}.
\end{aligned} \tag{12.6.10}$$

In doing this we just proved the following theorem: gradient descent with and without momentum for a convex quadratic function decomposes into coordinate-wise optimization in the direction of the eigenvectors of the quadratic matrix.

### Scalar Functions

Given the above result let's see what happens when we minimize the function $f(x) = \frac{\lambda}{2} x^2$. For gradient descent we have

$$x_{t+1} = x_t - \eta \lambda x_t = (1 - \eta \lambda) x_t. \tag{12.6.11}$$

Whenever $|1 - \eta \lambda| < 1$ this optimization converges at an exponential rate since after $t$ steps we have $x_t = (1 - \eta \lambda)^t x_0$. This shows how the rate of convergence improves initially as we increase the learning rate $\eta$ until $\eta \lambda = 1$. Beyond that things diverge and for $\eta \lambda > 2$ the optimization problem diverges.

```
lambdas = [0.1, 1, 10, 19]
eta = 0.1
d2l.set_figsize((6, 4))
for lam in lambdas:
    t = torch.arange(20).detach().numpy()
    d2l.plt.plot(t, (1 - eta * lam) ** t, label=f'lambda = {lam:.2f}')
d2l.plt.xlabel('time')
d2l.plt.legend();
```

To analyze convergence in the case of momentum we begin by rewriting the update equations in terms of two scalars: one for $x$ and one for velocity $v$. This yields:

$$\begin{bmatrix} v_{t+1} \\ x_{t+1} \end{bmatrix} =
\begin{bmatrix} \beta & \lambda \\ -\eta \beta & (1 - \eta \lambda) \end{bmatrix}
\begin{bmatrix} v_{t} \\ x_{t} \end{bmatrix} = \mathbf{R}(\beta, \eta, \lambda) \begin{bmatrix} v_{t} \\ x_{t} \end{bmatrix}. \tag{12.6.12}$$

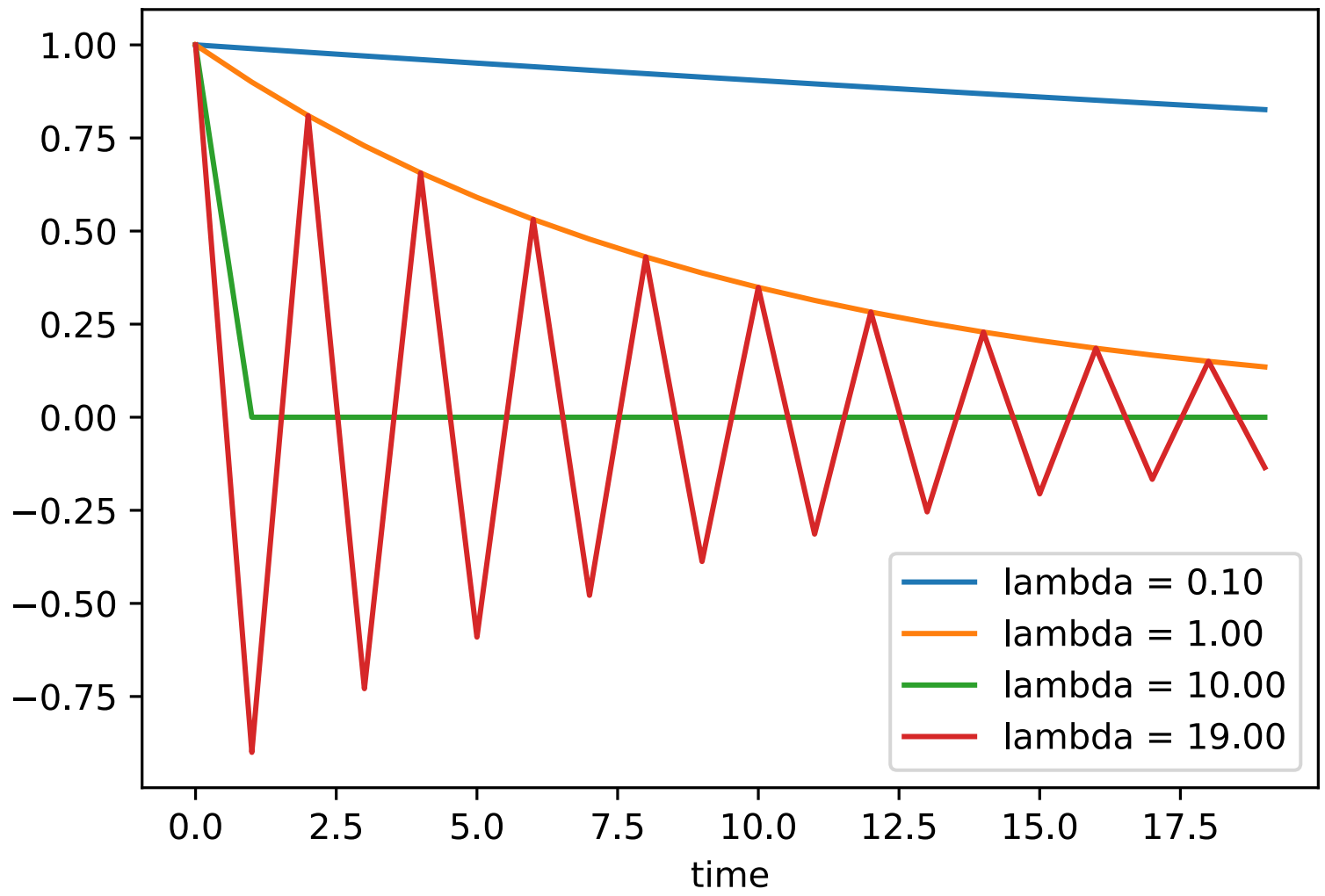

We used $\mathbf{R}$ to denote the $2 \times 2$ governing convergence behavior. After $t$ steps the initial choice $[v_0, x_0]$ becomes $\mathbf{R}(\beta, \eta, \lambda)^t [v_0, x_0]$. Hence, it is up to the eigenvalues of $\mathbf{R}$ to determine the speed of convergence. See the Distill post[175] of Goh (2017) for a great animation and Flammarion and Bach (2015) for a detailed analysis. One can show that $0 < \eta\lambda < 2 + 2\beta$ velocity converges. This is a larger range of feasible parameters when compared to $0 < \eta\lambda < 2$ for gradient descent. It also suggests that in general large values of $\beta$ are desirable. Further details require a fair amount of technical detail and we suggest that the interested reader consult the original publications.

## 12.6.4 Summary

- Momentum replaces gradients with a leaky average over past gradients. This accelerates convergence significantly.
- It is desirable for both noise-free gradient descent and (noisy) stochastic gradient descent.
- Momentum prevents stalling of the optimization process that is much more likely to occur for stochastic gradient descent.
- The effective number of gradients is given by $\frac{1}{1-\beta}$ due to exponentiated downweighting of past data.
- In the case of convex quadratic problems this can be analyzed explicitly in detail.
- Implementation is quite straightforward but it requires us to store an additional state vector (velocity $\mathbf{v}$).

## 12.6.5 Exercises

1. Use other combinations of momentum hyperparameters and learning rates and observe and analyze the different experimental results.
2. Try out gradient descent and momentum for a quadratic problem where you have multiple eigenvalues, i.e., $f(x) = \frac{1}{2} \sum_i \lambda_i x_i^2$, e.g., $\lambda_i = 2^{-i}$. Plot how the values of $x$ decrease for the initialization $x_i = 1$.
3. Derive minimum value and minimizer for $h(\mathbf{x}) = \frac{1}{2}\mathbf{x}^\top \mathbf{Q}\mathbf{x} + \mathbf{x}^\top \mathbf{c} + b$.
4. What changes when we perform stochastic gradient descent with momentum? What happens when we use minibatch stochastic gradient descent with momentum? Experiment with the parameters?

Discussions[176].

# 12.7 Adagrad

Let's begin by considering learning problems with features that occur infrequently.

## 12.7.1 Sparse Features and Learning Rates

Imagine that we are training a language model. To get good accuracy we typically want to decrease the learning rate as we keep on training, usually at a rate of $\mathcal{O}(t^{-\frac{1}{2}})$ or slower. Now consider a model training on sparse features, i.e., features that occur only infrequently. This is common for natural language, e.g., it is a lot less likely that we will see the word *preconditioning* than *learning*. However, it is also common in other areas such as computational advertising and personalized collaborative filtering. After all, there are many things that are of interest only for a small number of people.

Parameters associated with infrequent features only receive meaningful updates whenever these features occur. Given a decreasing learning rate we might end up in a situation where the parameters for common features converge rather quickly to their optimal values, whereas for infrequent features we are still short of observing them sufficiently frequently before their optimal values can be determined. In other words, the learning rate either decreases too slowly for frequent features or too quickly for infrequent ones.

A possible hack to redress this issue would be to count the number of times we see a particular feature and to use this as a clock for adjusting learning rates. That is, rather than choosing a learning rate of the form $\eta = \frac{\eta_0}{\sqrt{t+c}}$ we could use $\eta_i = \frac{\eta_0}{\sqrt{s(i,t)+c}}$. Here $s(i, t)$ counts the number of nonzeros for feature $i$ that we have observed up to time $t$. This is actually quite easy to implement at no meaningful overhead. However, it fails whenever we do not quite have sparsity but rather just data where the gradients are often very small and only rarely large. After all, it is unclear where one would draw the line between something that qualifies as an observed feature or not.

Adagrad by Duchi *et al.* (2011) addresses this by replacing the rather crude counter $s(i,t)$ by an aggregate of the squares of previously observed gradients. In particular, it uses $s(i,t+1) = s(i,t) + (\partial_i f(\mathbf{x}))^2$ as a means to adjust the learning rate. This has two benefits: first, we no longer need to decide just when a gradient is large enough. Second, it scales automatically with the magnitude of the gradients. Coordinates that routinely correspond to large gradients are scaled down significantly, whereas others with small gradients receive a much more gentle treatment. In practice this leads to a very effective optimization procedure for computational advertising and related problems. But this hides some of the additional benefits inherent in Adagrad that are best understood in the context of preconditioning.

## 12.7.2 Preconditioning

Convex optimization problems are good for analyzing the characteristics of algorithms. After all, for most nonconvex problems it is difficult to derive meaningful theoretical guarantees, but *intuition* and *insight* often carry over. Let's look at the problem of minimizing $f(\mathbf{x}) = \frac{1}{2}\mathbf{x}^\top \mathbf{Q}\mathbf{x} + \mathbf{c}^\top \mathbf{x} + b$.

As we saw in Section 12.6, it is possible to rewrite this problem in terms of its eigendecomposition $\mathbf{Q} = \mathbf{U}^\top \boldsymbol{\Lambda} \mathbf{U}$ to arrive at a much simplified problem where each coordinate can be solved individually:

$$f(\mathbf{x}) = \bar{f}(\bar{\mathbf{x}}) = \frac{1}{2}\bar{\mathbf{x}}^\top \boldsymbol{\Lambda} \bar{\mathbf{x}} + \bar{\mathbf{c}}^\top \bar{\mathbf{x}} + b. \tag{12.7.1}$$

Here we used $\bar{\mathbf{x}} = \mathbf{U}\mathbf{x}$ and consequently $\bar{\mathbf{c}} = \mathbf{U}\mathbf{c}$. The modified problem has as its minimizer $\bar{\mathbf{x}} = -\boldsymbol{\Lambda}^{-1}\bar{\mathbf{c}}$ and minimum value $-\frac{1}{2}\bar{\mathbf{c}}^\top \boldsymbol{\Lambda}^{-1}\bar{\mathbf{c}} + b$. This is much easier to compute since $\boldsymbol{\Lambda}$ is a diagonal matrix containing the eigenvalues of $\mathbf{Q}$.

If we perturb $\mathbf{c}$ slightly we would hope to find only slight changes in the minimizer of $f$. Unfortunately this is not the case. While slight changes in $\mathbf{c}$ lead to equally slight changes in $\bar{\mathbf{c}}$, this is not the case for the minimizer of $f$ (and of $\bar{f}$ respectively). Whenever the eigenvalues $\boldsymbol{\Lambda}_i$ are large we will see only small changes in $\bar{x}_i$ and in the minimum of $\bar{f}$. Conversely, for small $\boldsymbol{\Lambda}_i$ changes in $\bar{x}_i$ can be dramatic. The ratio between the largest and the smallest eigenvalue is called the condition number of an optimization problem.

$$\kappa = \frac{\boldsymbol{\Lambda}_1}{\boldsymbol{\Lambda}_d}. \tag{12.7.2}$$

If the condition number $\kappa$ is large, it is difficult to solve the optimization problem accurately. We need to ensure that we are careful in getting a large dynamic range of values right. Our analysis leads to an obvious, albeit somewhat naive question: couldn't we simply "fix" the problem by distorting the space such that all eigenvalues are 1. In theory this is quite easy: we only need the eigenvalues and eigenvectors of $\mathbf{Q}$ to rescale the problem from $\mathbf{x}$ to one in $\mathbf{z} \stackrel{\text{def}}{=} \boldsymbol{\Lambda}^{\frac{1}{2}}\mathbf{U}\mathbf{x}$. In the new coordinate system $\mathbf{x}^\top \mathbf{Q}\mathbf{x}$ could be simplified to $\|\mathbf{z}\|^2$. Alas, this is a rather impractical suggestion. Computing eigenvalues and eigenvectors is in general *much more* expensive than solving the actual problem.

While computing eigenvalues exactly might be expensive, guessing them and computing them even somewhat approximately may already be a lot better than not doing anything at all. In particular, we could use the diagonal entries of $\mathbf{Q}$ and rescale it accordingly. This is *much* cheaper than computing eigenvalues.

$$\tilde{\mathbf{Q}} = \mathrm{diag}^{-\frac{1}{2}}(\mathbf{Q}) \mathbf{Q} \mathrm{diag}^{-\frac{1}{2}}(\mathbf{Q}). \tag{12.7.3}$$

In this case we have $\tilde{\mathbf{Q}}_{ij} = \mathbf{Q}_{ij} / \sqrt{\mathbf{Q}_{ii} \mathbf{Q}_{jj}}$ and specifically $\tilde{\mathbf{Q}}_{ii} = 1$ for all $i$. In most cases this simplifies the condition number considerably. For instance, the cases we discussed previously, this would entirely eliminate the problem at hand since the problem is axis aligned.

Unfortunately we face yet another problem: in deep learning we typically do not even have access to the second derivative of the objective function: for $\mathbf{x} \in \mathbb{R}^d$ the second derivative even on a minibatch may require $O(d^2)$ space and work to compute, thus making it practically infeasible. The ingenious idea of Adagrad is to use a proxy for that elusive diagonal of the Hessian that is both relatively cheap to compute and effective—the magnitude of the gradient itself.

In order to see why this works, let's look at $\bar{f}(\bar{\mathbf{x}})$. We have that

$$\partial_{\bar{\mathbf{x}}} \bar{f}(\bar{\mathbf{x}}) = \boldsymbol{\Lambda} \bar{\mathbf{x}} + \bar{\mathbf{c}} = \boldsymbol{\Lambda} \left(\bar{\mathbf{x}} - \bar{\mathbf{x}}_0\right), \tag{12.7.4}$$

where $\bar{\mathbf{x}}_0$ is the minimizer of $\bar{f}$. Hence the magnitude of the gradient depends both on $\boldsymbol{\Lambda}$ and the distance from optimality. If $\bar{\mathbf{x}} - \bar{\mathbf{x}}_0$ did not change, this would be all that is needed. After all, in this case the magnitude of the gradient $\partial_{\bar{\mathbf{x}}} \bar{f}(\bar{\mathbf{x}})$ suffices. Since AdaGrad is a stochastic gradient descent algorithm, we will see gradients with nonzero variance even at optimality. As a result we can safely use the variance of the gradients as a cheap proxy for the scale of the Hessian. A thorough analysis is beyond the scope of this section (it would be several pages). We refer the reader to (Duchi *et al.*, 2011) for details.

## 12.7.3 The Algorithm

Let's formalize the discussion from above. We use the variable $\mathbf{s}_t$ to accumulate past gradient variance as follows.

$$\begin{aligned}
\mathbf{g}_t & = \partial_{\mathbf{w}} l(y_t, f(\mathbf{x}_t, \mathbf{w})), \\
\mathbf{s}_t & = \mathbf{s}_{t-1} + \mathbf{g}_t^2, \\
\mathbf{w}_t & = \mathbf{w}_{t-1} - \frac{\eta}{\sqrt{\mathbf{s}_t + \epsilon}} \cdot \mathbf{g}_t.
\end{aligned} \tag{12.7.5}$$

Here the operation are applied coordinate wise. That is, $\mathbf{v}^2$ has entries $v_i^2$. Likewise $\frac{1}{\sqrt{v}}$ has entries $\frac{1}{\sqrt{v_i}}$ and $\mathbf{u} \cdot \mathbf{v}$ has entries $u_i v_i$. As before $\eta$ is the learning rate and $\epsilon$ is an additive constant that ensures that we do not divide by 0. Last, we initialize $\mathbf{s}_0 = \mathbf{0}$.

Just like in the case of momentum we need to keep track of an auxiliary variable, in this case to allow for an individual learning rate per coordinate. This does not increase the cost of Adagrad significantly relative to SGD, simply since the main cost is typically to compute $l(y_t, f(\mathbf{x}_t, \mathbf{w}))$ and its derivative.

Note that accumulating squared gradients in $\mathbf{s}_t$ means that $\mathbf{s}_t$ grows essentially at linear rate (somewhat slower than linearly in practice, since the gradients initially diminish). This leads to an $O(t^{-\frac{1}{2}})$ learning rate, albeit adjusted on a per coordinate basis. For convex problems this is perfectly adequate. In deep learning, though, we might want to decrease the learning rate rather more slowly. This led to a number of Adagrad variants that we will discuss in the subsequent chapters. For now let's see how it behaves in a quadratic convex problem. We use the same problem as before:

$$f(\mathbf{x}) = 0.1x_1^2 + 2x_2^2. \tag{12.7.6}$$

We are going to implement Adagrad using the same learning rate previously, i.e., $\eta = 0.4$. As we can see, the iterative trajectory of the independent variable is smoother. However, due to the cumulative effect of $\boldsymbol{s}_t$, the learning rate continuously decays, so the independent variable does not move as much during later stages of iteration.

```
%matplotlib inline
import math
import torch
from d2l import torch as d2l
```

```
def adagrad_2d(x1, x2, s1, s2):
    eps = 1e-6
    g1, g2 = 0.2 * x1, 4 * x2
    s1 += g1 ** 2
    s2 += g2 ** 2
    x1 -= eta / math.sqrt(s1 + eps) * g1
    x2 -= eta / math.sqrt(s2 + eps) * g2
    return x1, x2, s1, s2

def f_2d(x1, x2):
    return 0.1 * x1 ** 2 + 2 * x2 ** 2

eta = 0.4
d2l.show_trace_2d(f_2d, d2l.train_2d(adagrad_2d))
```

```
epoch 20, x1: -2.382563, x2: -0.158591
```

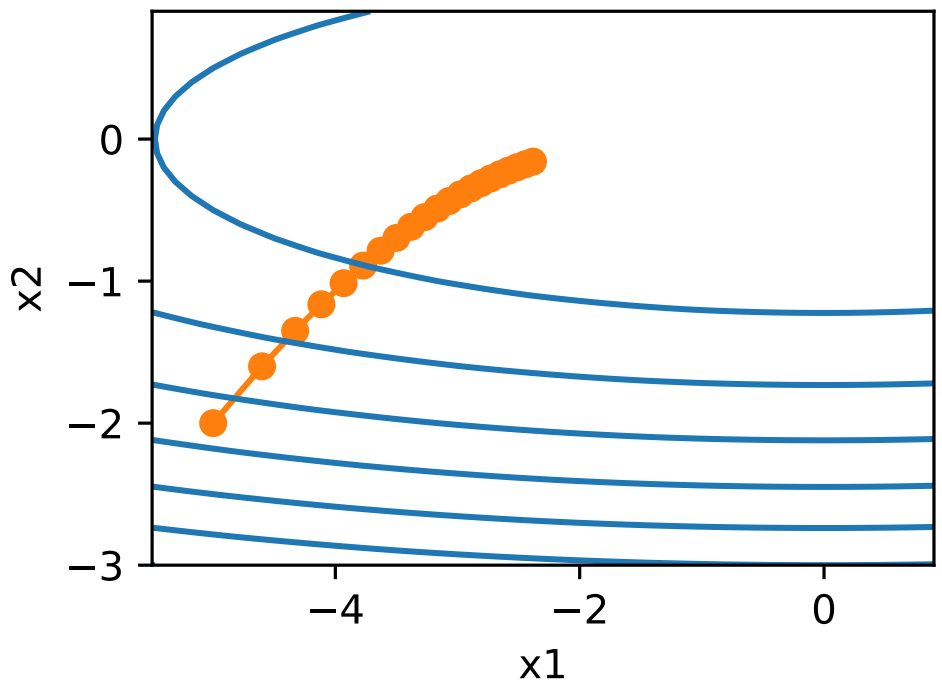

As we increase the learning rate to 2 we see much better behavior. This already indicates that the decrease in learning rate might be rather aggressive, even in the noise-free case and we need to ensure that parameters converge appropriately.

```
eta = 2
d2l.show_trace_2d(f_2d, d2l.train_2d(adagrad_2d))
```

```
epoch 20, x1: -0.002295, x2: -0.000000
```

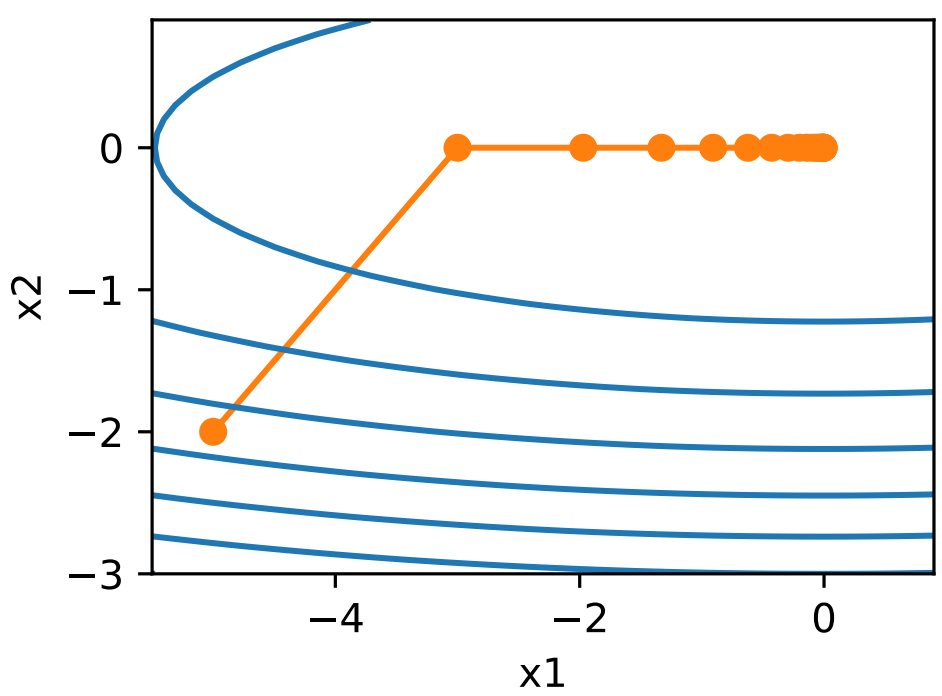

## 12.7.4 Implementation from Scratch

Just like the momentum method, Adagrad needs to maintain a state variable of the same shape as the parameters.

```
def init_adagrad_states(feature_dim):
    s_w = torch.zeros((feature_dim, 1))
    s_b = torch.zeros(1)
    return (s_w, s_b)

def adagrad(params, states, hyperparams):
    eps = 1e-6
    for p, s in zip(params, states):
        with torch.no_grad():
            s[:] += torch.square(p.grad)
            p[:] -= hyperparams['lr'] * p.grad / torch.sqrt(s + eps)
        p.grad.data.zero_()
```

Compared to the experiment in Section 12.5 we use a larger learning rate to train the model.

```
data_iter, feature_dim = d2l.get_data_ch11(batch_size=10)
d2l.train_ch11(adagrad, init_adagrad_states(feature_dim),
               {'lr': 0.1}, data_iter, feature_dim);
```

```
loss: 0.244, 0.232 sec/epoch
```

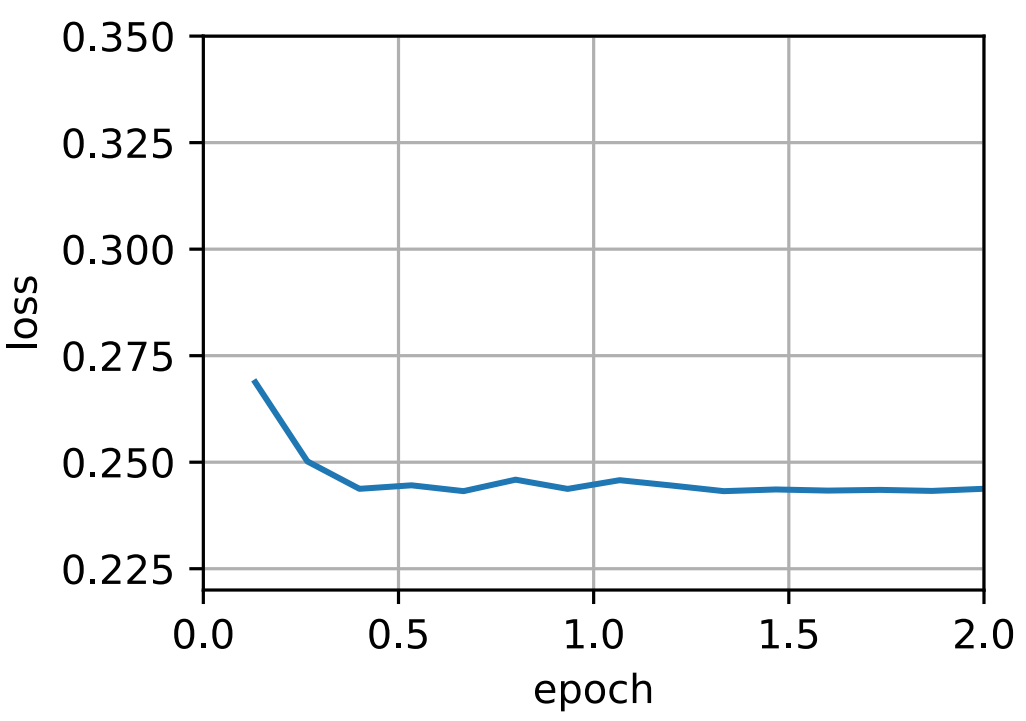

## 12.7.5 Concise Implementation

Using the `Trainer` instance of the algorithm `adagrad`, we can invoke the Adagrad algorithm in Gluon.

```
trainer = torch.optim.Adagrad
d2l.train_concise_ch11(trainer, {'lr': 0.1}, data_iter)
```

```
loss: 0.242, 0.155 sec/epoch
```

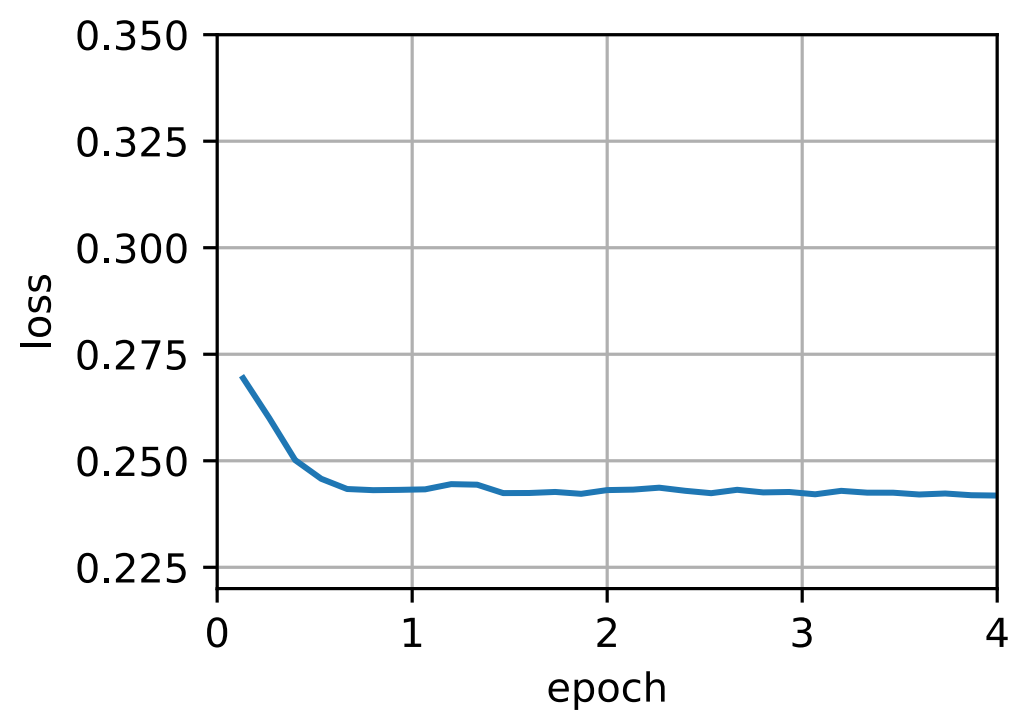

## 12.7.6 Summary

- Adagrad decreases the learning rate dynamically on a per-coordinate basis.
- It uses the magnitude of the gradient as a means of adjusting how quickly progress is

achieved - coordinates with large gradients are compensated with a smaller learning rate.

- Computing the exact second derivative is typically infeasible in deep learning problems due to memory and computational constraints. The gradient can be a useful proxy.
- If the optimization problem has a rather uneven structure Adagrad can help mitigate the distortion.
- Adagrad is particularly effective for sparse features where the learning rate needs to decrease more slowly for infrequently occurring terms.
- On deep learning problems Adagrad can sometimes be too aggressive in reducing learning rates. We will discuss strategies for mitigating this in the context of Section 12.10.

### 12.7.7 Exercises

1. Prove that for an orthogonal matrix $\mathbf{U}$ and a vector $\mathbf{c}$ the following holds: $\|\mathbf{c} - \mathbf{ffi}\|_2 = \|\mathbf{U}\mathbf{c} - \mathbf{U}\mathbf{ffi}\|_2$. Why does this mean that the magnitude of perturbations does not change after an orthogonal change of variables?
2. Try out Adagrad for $f(\mathbf{x}) = 0.1x_1^2 + 2x_2^2$ and also for the objective function was rotated by 45 degrees, i.e., $f(\mathbf{x}) = 0.1(x_1 + x_2)^2 + 2(x_1 - x_2)^2$. Does it behave differently?
3. Prove Gerschgorin's circle theorem[177] which states that eigenvalues $\lambda_i$ of a matrix $\mathbf{M}$ satisfy $|\lambda_i - \mathbf{M}_{jj}| \leq \sum_{k \neq j} |\mathbf{M}_{jk}|$ for at least one choice of $j$.
4. What does Gerschgorin's theorem tell us about the eigenvalues of the diagonally preconditioned matrix $\mathrm{diag}^{-\frac{1}{2}}(\mathbf{M})\mathbf{M}\mathrm{diag}^{-\frac{1}{2}}(\mathbf{M})$?
5. Try out Adagrad for a proper deep network, such as Section 7.6 when applied to Fashion-MNIST.
6. How would you need to modify Adagrad to achieve a less aggressive decay in learning rate?

Discussions[178].

## 12.8 RMSProp

One of the key issues in Section 12.7 is that the learning rate decreases at a predefined schedule of effectively $O(t^{-\frac{1}{2}})$. While this is generally appropriate for convex problems, it might not be ideal for nonconvex ones, such as those encountered in deep learning. Yet, the coordinate-wise adaptivity of Adagrad is highly desirable as a preconditioner.

Tieleman and Hinton (2012) proposed the RMSProp algorithm as a simple fix to decouple rate scheduling from coordinate-adaptive learning rates. The issue is that Adagrad accumulates the squares of the gradient $\mathbf{g}_t$ into a state vector $\mathbf{s}_t = \mathbf{s}_{t-1} + \mathbf{g}_t^2$. As a result $\mathbf{s}_t$ keeps on growing without bound due to the lack of normalization, essentially linearly as the algorithm converges.

One way of fixing this problem would be to use $\mathbf{s}_t / t$. For reasonable distributions of $\mathbf{g}_t$ this will converge. Unfortunately it might take a very long time until the limit behavior starts to matter since the procedure remembers the full trajectory of values. An alternative is to use a leaky average in the same way we used in the momentum method, i.e., $\mathbf{s}_t \leftarrow \gamma \mathbf{s}_{t-1} + (1-\gamma) \mathbf{g}_t^2$ for some parameter $\gamma > 0$. Keeping all other parts unchanged yields RMSProp.

## 12.8.1 The Algorithm

Let's write out the equations in detail.

$$\begin{aligned}
\mathbf{s}_t & \leftarrow \gamma \mathbf{s}_{t-1} + (1 - \gamma) \mathbf{g}_t^2, \\
\mathbf{x}_t & \leftarrow \mathbf{x}_{t-1} - \frac{\eta}{\sqrt{\mathbf{s}_t + \epsilon}} \odot \mathbf{g}_t.
\end{aligned} \tag{12.8.1}$$

The constant $\epsilon > 0$ is typically set to $10^{-6}$ to ensure that we do not suffer from division by zero or overly large step sizes. Given this expansion we are now free to control the learning rate $\eta$ independently of the scaling that is applied on a per-coordinate basis. In terms of leaky averages we can apply the same reasoning as previously applied in the case of the momentum method. Expanding the definition of $\mathbf{s}_t$ yields

$$\begin{aligned}
\mathbf{s}_t & = (1 - \gamma) \mathbf{g}_t^2 + \gamma \mathbf{s}_{t-1} \\
& = (1 - \gamma) \left(\mathbf{g}_t^2 + \gamma \mathbf{g}_{t-1}^2 + \gamma^2 \mathbf{g}_{t-2} + \ldots, \right).
\end{aligned} \tag{12.8.2}$$

As before in Section 12.6 we use $1 + \gamma + \gamma^2 + \ldots, = \frac{1}{1-\gamma}$. Hence the sum of weights is normalized to 1 with a half-life time of an observation of $\gamma^{-1}$. Let's visualize the weights for the past 40 time steps for various choices of $\gamma$.

```
import math
import torch
from d2l import torch as d2l
```

```
d2l.set_figsize()
gammas = [0.95, 0.9, 0.8, 0.7]
for gamma in gammas:
    x = torch.arange(40).detach().numpy()
    d2l.plt.plot(x, (1-gamma) * gamma ** x, label=f'gamma = {gamma:.2f}')
d2l.plt.xlabel('time');
```

## 12.8.2 Implementation from Scratch

As before we use the quadratic function $f(\mathbf{x}) = 0.1x_1^2 + 2x_2^2$ to observe the trajectory of RMSProp. Recall that in Section 12.7, when we used Adagrad with a learning rate of 0.4, the variables moved only very slowly in the later stages of the algorithm since the learning rate decreased too quickly. Since $\eta$ is controlled separately this does not happen with RMSProp.

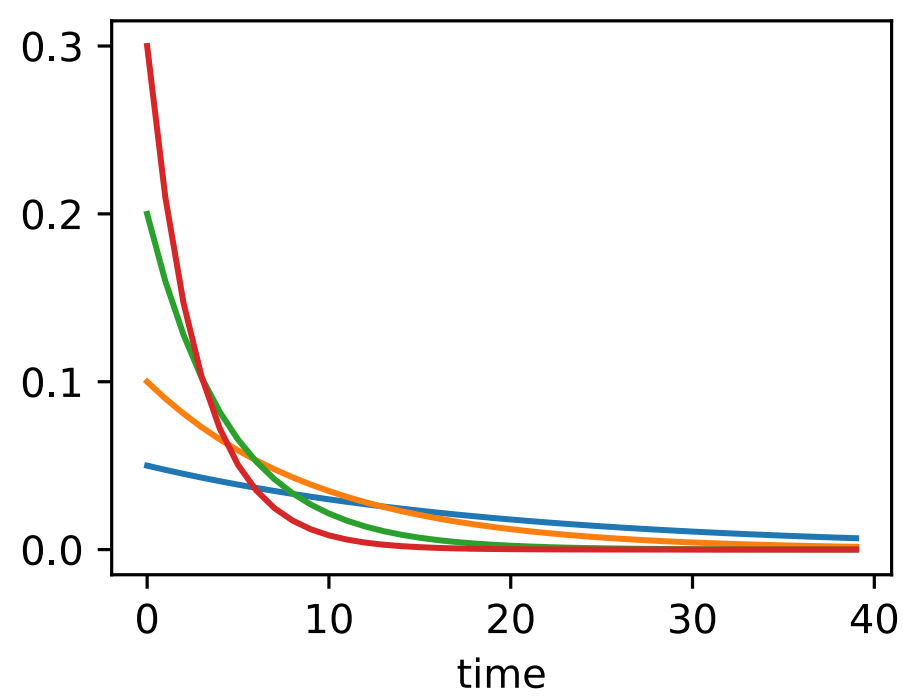

```
def rmsprop_2d(x1, x2, s1, s2):
    g1, g2, eps = 0.2 * x1, 4 * x2, 1e-6
    s1 = gamma * s1 + (1 - gamma) * g1 ** 2
    s2 = gamma * s2 + (1 - gamma) * g2 ** 2
    x1 -= eta / math.sqrt(s1 + eps) * g1
    x2 -= eta / math.sqrt(s2 + eps) * g2
    return x1, x2, s1, s2

def f_2d(x1, x2):
    return 0.1 * x1 ** 2 + 2 * x2 ** 2

eta, gamma = 0.4, 0.9
d2l.show_trace_2d(f_2d, d2l.train_2d(rmsprop_2d))
```

```
epoch 20, x1: -0.010599, x2: 0.000000
```

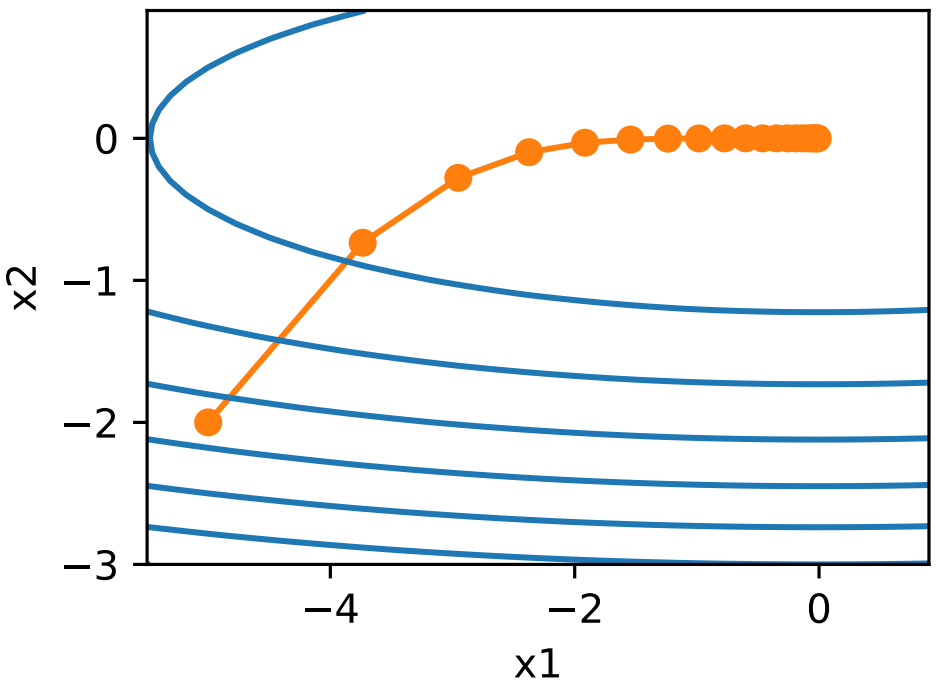

Next, we implement RMSProp to be used in a deep network. This is equally straightforward.

```
def init_rmsprop_states(feature_dim):
    s_w = torch.zeros((feature_dim, 1))
    s_b = torch.zeros(1)
    return (s_w, s_b)
```

```
def rmsprop(params, states, hyperparams):
    gamma, eps = hyperparams['gamma'], 1e-6
    for p, s in zip(params, states):
        with torch.no_grad():
            s[:] = gamma * s + (1 - gamma) * torch.square(p.grad)
            p[:] -= hyperparams['lr'] * p.grad / torch.sqrt(s + eps)
        p.grad.data.zero_()
```

We set the initial learning rate to 0.01 and the weighting term $\gamma$ to 0.9. That is, s aggregates on average over the past $1/(1 - \gamma) = 10$ observations of the square gradient.

```
data_iter, feature_dim = d2l.get_data_ch11(batch_size=10)
d2l.train_ch11(rmsprop, init_rmsprop_states(feature_dim),
               {'lr': 0.01, 'gamma': 0.9}, data_iter, feature_dim);
```

```
loss: 0.243, 0.177 sec/epoch
```

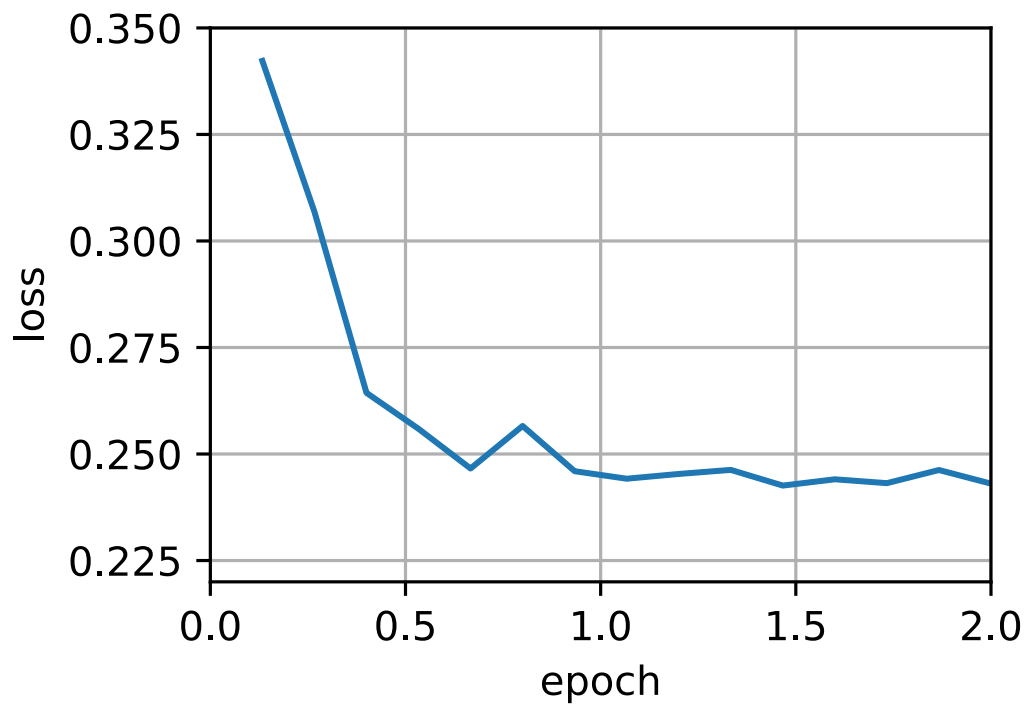

## 12.8.3 Concise Implementation

Since RMSProp is a rather popular algorithm it is also available in the `Trainer` instance. All we need to do is instantiate it using an algorithm named `rmsprop`, assigning $\gamma$ to the parameter `gamma1`.

```
trainer = torch.optim.RMSprop
d2l.train_concise_ch11(trainer, {'lr': 0.01, 'alpha': 0.9},
                       data_iter)
```

```
loss: 0.243, 0.138 sec/epoch
```

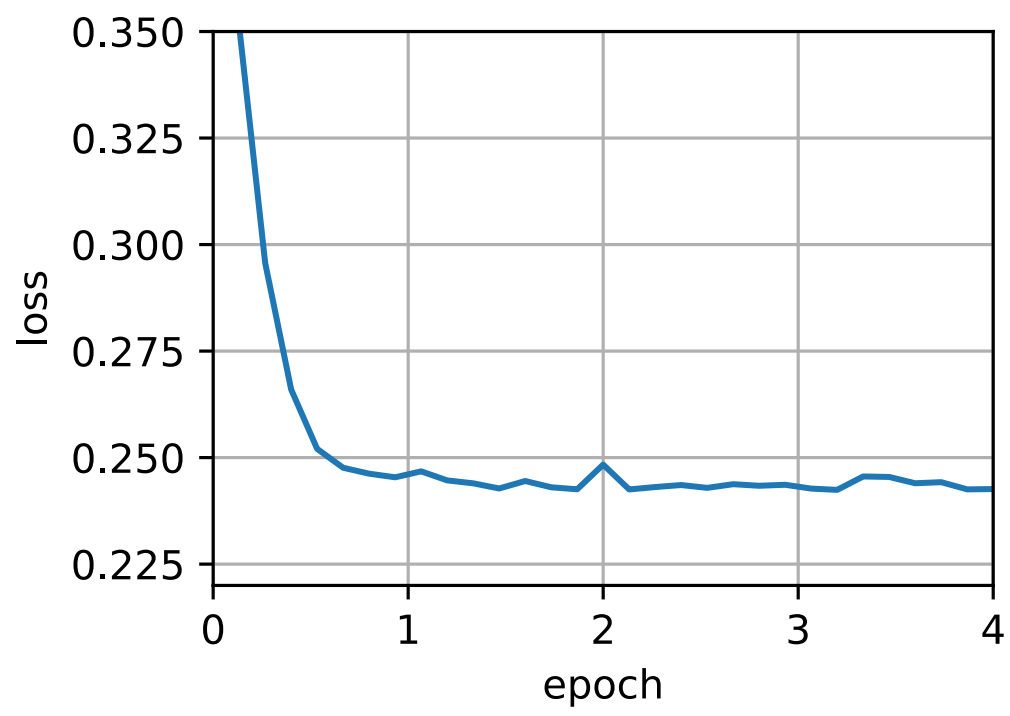

### 12.8.4 Summary

- RMSProp is very similar to Adagrad insofar as both use the square of the gradient to scale coefficients.
- RMSProp shares with momentum the leaky averaging. However, RMSProp uses the technique to adjust the coefficient-wise preconditioner.
- The learning rate needs to be scheduled by the experimenter in practice.
- The coefficient $\gamma$ determines how long the history is when adjusting the per-coordinate scale.

### 12.8.5 Exercises

1. What happens experimentally if we set $\gamma = 1$? Why?
2. Rotate the optimization problem to minimize $f(\mathbf{x}) = 0.1(x_1 + x_2)^2 + 2(x_1 - x_2)^2$. What happens to the convergence?
3. Try out what happens to RMSProp on a real machine learning problem, such as training on Fashion-MNIST. Experiment with different choices for adjusting the learning rate.
4. Would you want to adjust $\gamma$ as optimization progresses? How sensitive is RMSProp to this?

Discussions[179].

# 12.9 Adadelta

Adadelta is yet another variant of AdaGrad (Section 12.7). The main difference lies in the fact that it decreases the amount by which the learning rate is adaptive to coordinates. Moreover, traditionally it referred to as not having a learning rate since it uses the amount of change itself as calibration for future change. The algorithm was proposed in Zeiler (2012). It is fairly straightforward, given the discussion of previous algorithms so far.

## 12.9.1 The Algorithm

In a nutshell, Adadelta uses two state variables, $\mathbf{s}_t$ to store a leaky average of the second moment of the gradient and $\Delta\mathbf{x}_t$ to store a leaky average of the second moment of the change of parameters in the model itself. Note that we use the original notation and naming of the authors for compatibility with other publications and implementations (there is no other real reason why one should use different Greek variables to indicate a parameter serving the same purpose in momentum, Adagrad, RMSProp, and Adadelta).

Here are the technical details of Adadelta. Given the parameter du jour is $\rho$, we obtain the following leaky updates similarly to Section 12.8:

$$\mathbf{s}_t = \rho \mathbf{s}_{t-1} + (1-\rho)\mathbf{g}_t^2. \tag{12.9.1}$$

The difference to Section 12.8 is that we perform updates with the rescaled gradient $\mathbf{g}_t'$, i.e.,

$$\mathbf{x}_t = \mathbf{x}_{t-1} - \mathbf{g}_t'. \tag{12.9.2}$$

So what is the rescaled gradient $\mathbf{g}_t'$? We can calculate it as follows:

$$\mathbf{g}_t' = \frac{\sqrt{\Delta\mathbf{x}_{t-1} + \epsilon}}{\sqrt{\mathbf{s}_t + \epsilon}} \odot \mathbf{g}_t, \tag{12.9.3}$$

where $\Delta\mathbf{x}_{t-1}$ is the leaky average of the squared rescaled gradients $\mathbf{g}_t'$. We initialize $\Delta\mathbf{x}_0$ to be $0$ and update it at each step with $\mathbf{g}_t'$, i.e.,

$$\Delta\mathbf{x}_t = \rho\Delta\mathbf{x}_{t-1} + (1-\rho){\mathbf{g}_t'}^2, \tag{12.9.4}$$

and $\epsilon$ (a small value such as $10^{-5}$) is added to maintain numerical stability.

## 12.9.2 Implementation

Adadelta needs to maintain two state variables for each variable, $\mathbf{s}_t$ and $\Delta\mathbf{x}_t$. This yields the following implementation.

```
%matplotlib inline
import torch
from d2l import torch as d2l

def init_adadelta_states(feature_dim):
    s_w, s_b = torch.zeros((feature_dim, 1)), torch.zeros(1)
    delta_w, delta_b = torch.zeros((feature_dim, 1)), torch.zeros(1)
    return ((s_w, delta_w), (s_b, delta_b))

def adadelta(params, states, hyperparams):
    rho, eps = hyperparams['rho'], 1e-5
    for p, (s, delta) in zip(params, states):
        with torch.no_grad():
            # In-place updates via [:]
            s[:] = rho * s + (1 - rho) * torch.square(p.grad)
            g = (torch.sqrt(delta + eps) / torch.sqrt(s + eps)) * p.grad
            p[:] -= g
            delta[:] = rho * delta + (1 - rho) * g * g
        p.grad.data.zero_()
```

Choosing $\rho = 0.9$ amounts to a half-life time of 10 for each parameter update. This tends to work quite well. We get the following behavior.

```
data_iter, feature_dim = d2l.get_data_ch11(batch_size=10)
d2l.train_ch11(adadelta, init_adadelta_states(feature_dim),
               {'rho': 0.9}, data_iter, feature_dim);
```

```
loss: 0.244, 0.221 sec/epoch
```

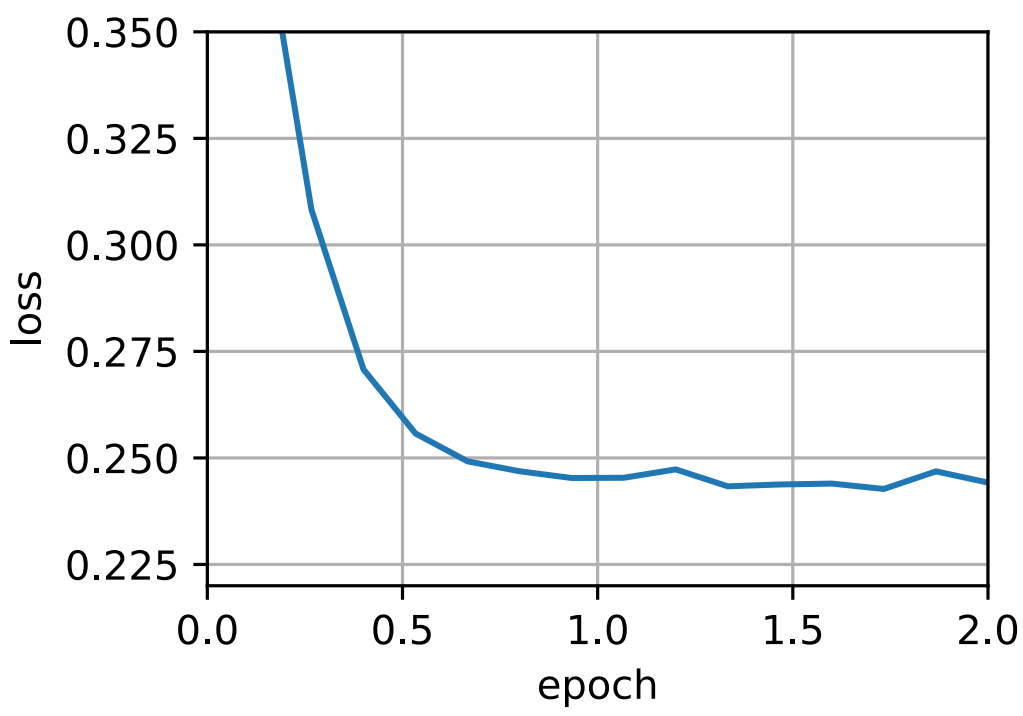

For a concise implementation we simply use the Adadelta algorithm from high-level APIs. This yields the following one-liner for a much more compact invocation.

```
trainer = torch.optim.Adadelta
d2l.train_concise_ch11(trainer, {'rho': 0.9}, data_iter)
```

```
loss: 0.243, 0.164 sec/epoch
```

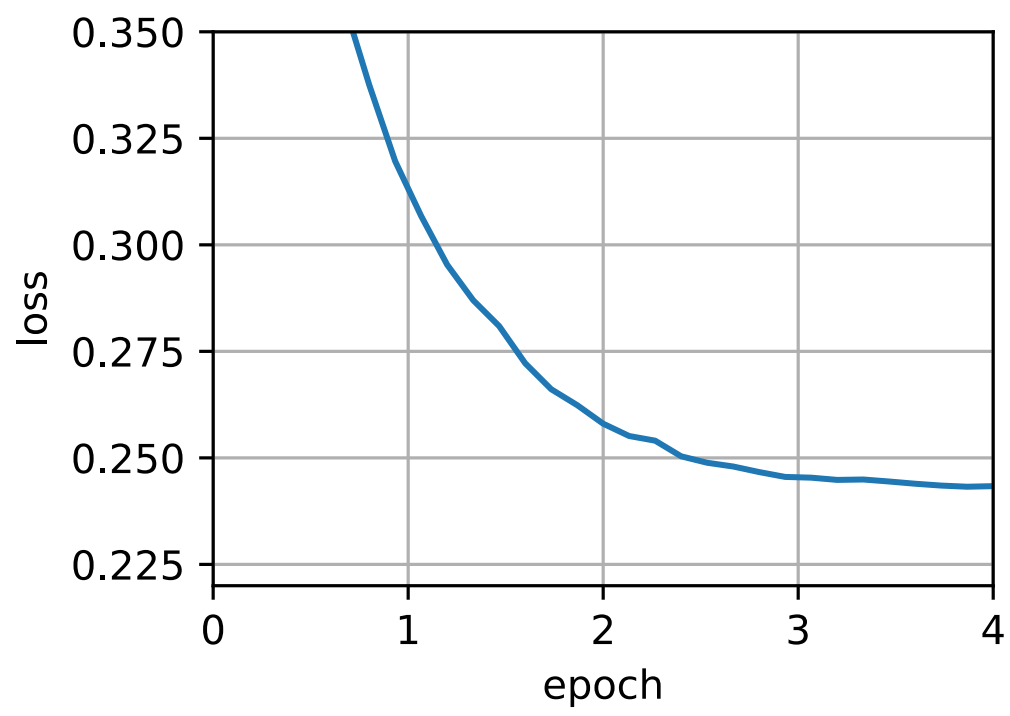

## 12.9.3 Summary

- Adadelta has no learning rate parameter. Instead, it uses the rate of change in the parameters itself to adapt the learning rate.
- Adadelta requires two state variables to store the second moments of gradient and the change in parameters.
- Adadelta uses leaky averages to keep a running estimate of the appropriate statistics.

## 12.9.4 Exercises

1. Adjust the value of $\rho$. What happens?
2. Show how to implement the algorithm without the use of $\mathbf{g}'_t$. Why might this be a good idea?
3. Is Adadelta really learning rate free? Could you find optimization problems that break Adadelta?
4. Compare Adadelta to Adagrad and RMS prop to discuss their convergence behavior.

Discussions[180].

# 12.10 Adam

In the discussions leading up to this section we encountered a number of techniques for efficient optimization. Let's recap them in detail here:

- We saw that Section 12.4 is more effective than Gradient Descent when solving optimization problems, e.g., due to its inherent resilience to redundant data.
- We saw that Section 12.5 affords significant additional efficiency arising from vectorization, using larger sets of observations in one minibatch. This is the key to efficient multi-machine, multi-GPU and overall parallel processing.
- Section 12.6 added a mechanism for aggregating a history of past gradients to accelerate convergence.
- Section 12.7 used per-coordinate scaling to allow for a computationally efficient preconditioner.
- Section 12.8 decoupled per-coordinate scaling from a learning rate adjustment.

Adam (Kingma and Ba, 2014) combines all these techniques into one efficient learning algorithm. As expected, this is an algorithm that has become rather popular as one of the more robust and effective optimization algorithms to use in deep learning. It is not without issues, though. In particular, (Reddi *et al.*, 2019) show that there are situations where Adam can diverge due to poor variance control. In a follow-up work Zaheer *et al.* (2018) proposed a hotfix to Adam, called Yogi which addresses these issues. More on this later. For now let's review the Adam algorithm.

## 12.10.1 The Algorithm

One of the key components of Adam is that it uses exponential weighted moving averages (also known as leaky averaging) to obtain an estimate of both the momentum and also the second moment of the gradient. That is, it uses the state variables

$$\begin{aligned}\mathbf{v}_t &\leftarrow \beta_1 \mathbf{v}_{t-1} + (1 - \beta_1) \mathbf{g}_t, \\ \mathbf{s}_t &\leftarrow \beta_2 \mathbf{s}_{t-1} + (1 - \beta_2) \mathbf{g}_t^2.\end{aligned} \tag{12.10.1}$$

Here $\beta_1$ and $\beta_2$ are nonnegative weighting parameters. Common choices for them are $\beta_1 = 0.9$ and $\beta_2 = 0.999$. That is, the variance estimate moves *much more slowly* than the momentum term. Note that if we initialize $\mathbf{v}_0 = \mathbf{s}_0 = 0$ we have a significant amount of bias initially towards smaller values. This can be addressed by using the fact that $\sum_{i=0}^{t-1} \beta^i = \frac{1 - \beta^t}{1 - \beta}$ to re-normalize terms. Correspondingly the normalized state variables are given by

$$\hat{\mathbf{v}}_t = \frac{\mathbf{v}_t}{1 - \beta_1^t} \text{ and } \hat{\mathbf{s}}_t = \frac{\mathbf{s}_t}{1 - \beta_2^t}. \tag{12.10.2}$$

Armed with the proper estimates we can now write out the update equations. First, we rescale the gradient in a manner very much akin to that of RMSProp to obtain

$$\mathbf{g}_t' = \frac{\eta \hat{\mathbf{v}}_t}{\sqrt{\hat{\mathbf{s}}_t} + \epsilon}. \tag{12.10.3}$$

Unlike RMSProp our update uses the momentum $\hat{\mathbf{v}}_t$ rather than the gradient itself. Moreover, there is a slight cosmetic difference as the rescaling happens using $\frac{1}{\sqrt{\hat{\mathbf{s}}_t}+\epsilon}$ instead of $\frac{1}{\sqrt{\hat{\mathbf{s}}_t+\epsilon}}$. The former works arguably slightly better in practice, hence the deviation from RMSProp. Typically we pick $\epsilon = 10^{-6}$ for a good trade-off between numerical stability and fidelity.

Now we have all the pieces in place to compute updates. This is slightly anticlimactic and we have a simple update of the form

$$\mathbf{x}_t \leftarrow \mathbf{x}_{t-1} - \mathbf{g}_t'. \tag{12.10.4}$$

Reviewing the design of Adam its inspiration is clear. Momentum and scale are clearly visible in the state variables. Their rather peculiar definition forces us to debias terms (this could be fixed by a slightly different initialization and update condition). Second, the combination of both terms is pretty straightforward, given RMSProp. Last, the explicit learning rate $\eta$ allows us to control the step length to address issues of convergence.

## 12.10.2 Implementation

Implementing Adam from scratch is not very daunting. For convenience we store the time step counter $t$ in the `hyperparams` dictionary. Beyond that all is straightforward.

```
%matplotlib inline
import torch
from d2l import torch as d2l

def init_adam_states(feature_dim):
    v_w, v_b = torch.zeros((feature_dim, 1)), torch.zeros(1)
    s_w, s_b = torch.zeros((feature_dim, 1)), torch.zeros(1)
    return ((v_w, s_w), (v_b, s_b))

def adam(params, states, hyperparams):
    beta1, beta2, eps = 0.9, 0.999, 1e-6
    for p, (v, s) in zip(params, states):
        with torch.no_grad():
            v[:] = beta1 * v + (1 - beta1) * p.grad
            s[:] = beta2 * s + (1 - beta2) * torch.square(p.grad)
            v_bias_corr = v / (1 - beta1 ** hyperparams['t'])
            s_bias_corr = s / (1 - beta2 ** hyperparams['t'])
            p[:] -= hyperparams['lr'] * v_bias_corr / (torch.sqrt(s_bias_corr)
                                                       + eps)
        p.grad.data.zero_()
    hyperparams['t'] += 1
```

We are ready to use Adam to train the model. We use a learning rate of $\eta = 0.01$.

```
data_iter, feature_dim = d2l.get_data_ch11(batch_size=10)
d2l.train_ch11(adam, init_adam_states(feature_dim),
               {'lr': 0.01, 't': 1}, data_iter, feature_dim);
```

```
loss: 0.246, 0.224 sec/epoch
```

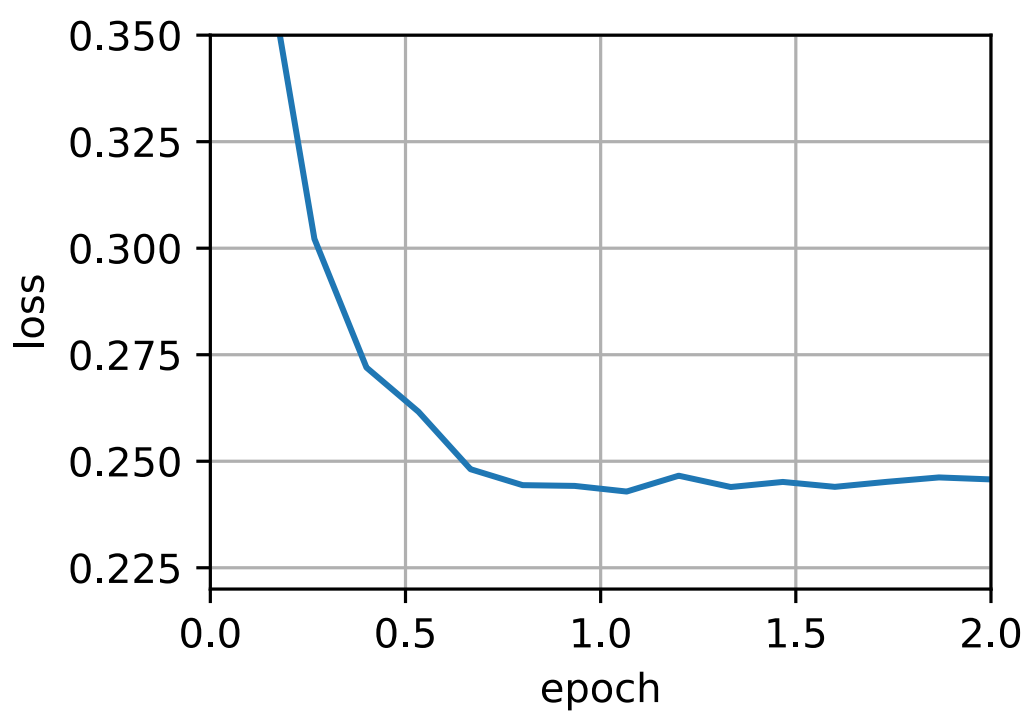

A more concise implementation is straightforward since `adam` is one of the algorithms provided as part of the Gluon `trainer` optimization library. Hence we only need to pass configuration parameters for an implementation in Gluon.

```
trainer = torch.optim.Adam
d2l.train_concise_ch11(trainer, {'lr': 0.01}, data_iter)
```

```
loss: 0.243, 0.182 sec/epoch
```

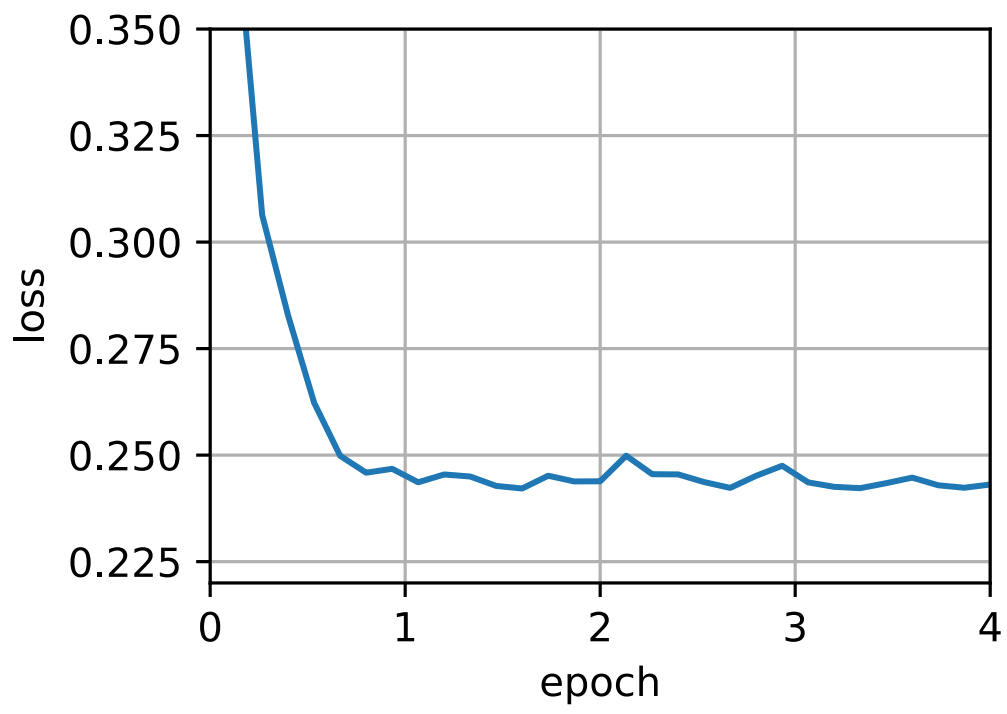

### 12.10.3 Yogi

One of the problems of Adam is that it can fail to converge even in convex settings when the second moment estimate in $\mathbf{s}_t$ blows up. As a fix Zaheer *et al.* (2018) proposed a refined update (and initialization) for $\mathbf{s}_t$. To understand what's going on, let's rewrite the Adam update as follows:

$$\mathbf{s}_t \leftarrow \mathbf{s}_{t-1} + (1 - \beta_2) \left(\mathbf{g}_t^2 - \mathbf{s}_{t-1}\right). \tag{12.10.5}$$

Whenever $\mathbf{g}_t^2$ has high variance or updates are sparse, $\mathbf{s}_t$ might forget past values too quickly. A possible fix for this is to replace $\mathbf{g}_t^2 - \mathbf{s}_{t-1}$ by $\mathbf{g}_t^2 \odot \mathop{\mathrm{sgn}}(\mathbf{g}_t^2 - \mathbf{s}_{t-1})$. Now the magnitude of the update no longer depends on the amount of deviation. This yields the Yogi updates

$$\mathbf{s}_t \leftarrow \mathbf{s}_{t-1} + (1 - \beta_2) \mathbf{g}_t^2 \odot \mathop{\mathrm{sgn}}(\mathbf{g}_t^2 - \mathbf{s}_{t-1}). \tag{12.10.6}$$

The authors furthermore advise to initialize the momentum on a larger initial batch rather than just initial pointwise estimate. We omit the details since they are not material to the discussion and since even without this convergence remains pretty good.

```
def yogi(params, states, hyperparams):
    beta1, beta2, eps = 0.9, 0.999, 1e-3
    for p, (v, s) in zip(params, states):
        with torch.no_grad():
            v[:] = beta1 * v + (1 - beta1) * p.grad
            s[:] = s + (1 - beta2) * torch.sign(
                torch.square(p.grad) - s) * torch.square(p.grad)
            v_bias_corr = v / (1 - beta1 ** hyperparams['t'])
            s_bias_corr = s / (1 - beta2 ** hyperparams['t'])
            p[:] -= hyperparams['lr'] * v_bias_corr / (torch.sqrt(s_bias_corr)
                                                       + eps)
        p.grad.data.zero_()
    hyperparams['t'] += 1

data_iter, feature_dim = d2l.get_data_ch11(batch_size=10)
d2l.train_ch11(yogi, init_adam_states(feature_dim),
               {'lr': 0.01, 't': 1}, data_iter, feature_dim);
```

```
loss: 0.245, 0.207 sec/epoch
```

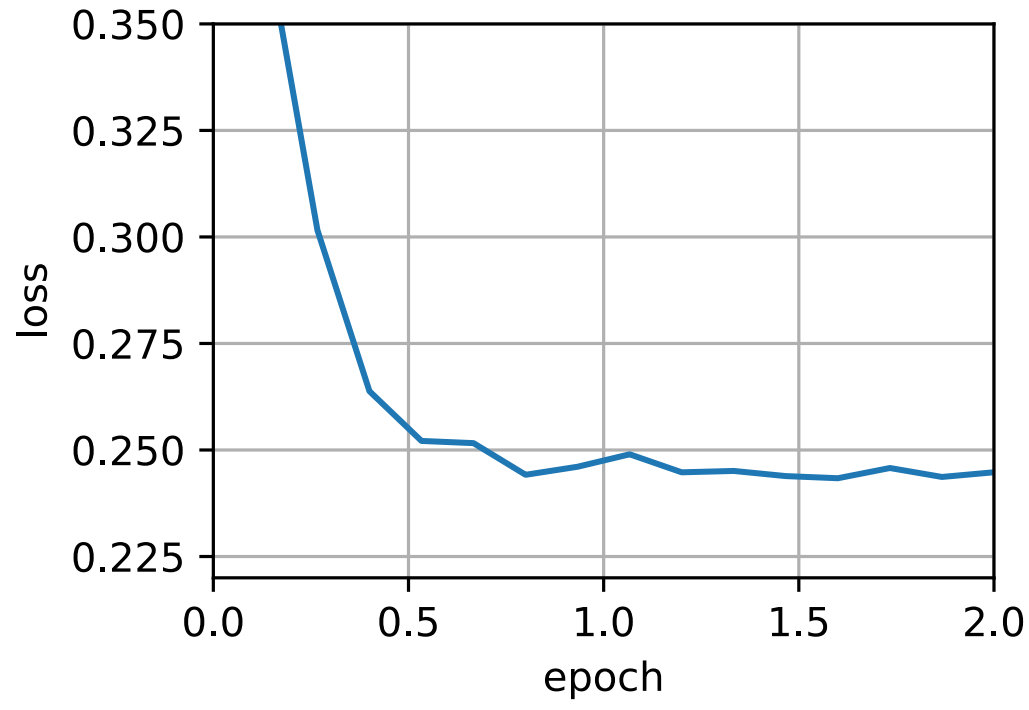

### 12.10.4 Summary

- Adam combines features of many optimization algorithms into a fairly robust update rule.
- Created on the basis of RMSProp, Adam also uses EWMA on the minibatch stochastic gradient.
- Adam uses bias correction to adjust for a slow startup when estimating momentum and a second moment.
- For gradients with significant variance we may encounter issues with convergence. They can be amended by using larger minibatches or by switching to an improved estimate for $\mathbf{s}_t$. Yogi offers such an alternative.

### 12.10.5 Exercises

1. Adjust the learning rate and observe and analyze the experimental results.
2. Can you rewrite momentum and second moment updates such that it does not require bias correction?
3. Why do you need to reduce the learning rate $\eta$ as we converge?
4. Try to construct a case for which Adam diverges and Yogi converges?

Discussions[181].

## 12.11 Learning Rate Scheduling

So far we primarily focused on optimization *algorithms* for how to update the weight vectors rather than on the *rate* at which they are being updated. Nonetheless, adjusting the learning rate is often just as important as the actual algorithm. There are a number of aspects to consider:

- Most obviously the *magnitude* of the learning rate matters. If it is too large, optimization diverges, if it is too small, it takes too long to train or we end up with a suboptimal result. We saw previously that the condition number of the problem matters (see e.g., Section 12.6 for details). Intuitively it is the ratio of the amount of change in the least sensitive direction vs. the most sensitive one.
- Secondly, the rate of decay is just as important. If the learning rate remains large we may simply end up bouncing around the minimum and thus not reach optimality. Section 12.5 discussed this in some detail and we analyzed performance guarantees in Section 12.4. In short, we want the rate to decay, but probably more slowly than $O(t^{-\frac{1}{2}})$ which would be a good choice for convex problems.

- Another aspect that is equally important is *initialization*. This pertains both to how the parameters are set initially (review Section 5.4 for details) and also how they evolve initially. This goes under the moniker of *warmup*, i.e., how rapidly we start moving towards the solution initially. Large steps in the beginning might not be beneficial, in particular since the initial set of parameters is random. The initial update directions might be quite meaningless, too.
- Lastly, there are a number of optimization variants that perform cyclical learning rate adjustment. This is beyond the scope of the current chapter. We recommend the reader to review details in Izmailov *et al.* (2018), e.g., how to obtain better solutions by averaging over an entire *path* of parameters.

Given the fact that there is a lot of detail needed to manage learning rates, most deep learning frameworks have tools to deal with this automatically. In the current chapter we will review the effects that different schedules have on accuracy and also show how this can be managed efficiently via a *learning rate scheduler*.

## 12.11.1 Toy Problem

We begin with a toy problem that is cheap enough to compute easily, yet sufficiently nontrivial to illustrate some of the key aspects. For that we pick a slightly modernized version of LeNet (`relu` instead of `sigmoid` activation, MaxPooling rather than AveragePooling), as applied to Fashion-MNIST. Moreover, we hybridize the network for performance. Since most of the code is standard we just introduce the basics without further detailed discussion. See Chapter 7 for a refresher as needed.

```
%matplotlib inline
import math
import torch
from torch import nn
from torch.optim import lr_scheduler
from d2l import torch as d2l

def net_fn():
    model = nn.Sequential(
        nn.Conv2d(1, 6, kernel_size=5, padding=2), nn.ReLU(),
        nn.MaxPool2d(kernel_size=2, stride=2),
        nn.Conv2d(6, 16, kernel_size=5), nn.ReLU(),
        nn.MaxPool2d(kernel_size=2, stride=2),
        nn.Flatten(),
        nn.Linear(16 * 5 * 5, 120), nn.ReLU(),
        nn.Linear(120, 84), nn.ReLU(),
        nn.Linear(84, 10))

    return model

loss = nn.CrossEntropyLoss()
device = d2l.try_gpu()
```

```
batch_size = 256
train_iter, test_iter = d2l.load_data_fashion_mnist(batch_size=batch_size)

# The code is almost identical to `d2l.train_ch6` defined in the
# lenet section of chapter convolutional neural networks
def train(net, train_iter, test_iter, num_epochs, loss, trainer, device,
          scheduler=None):
    net.to(device)
    animator = d2l.Animator(xlabel='epoch', xlim=[0, num_epochs],
                            legend=['train loss', 'train acc', 'test acc'])

    for epoch in range(num_epochs):
        metric = d2l.Accumulator(3)  # train_loss, train_acc, num_examples
        for i, (X, y) in enumerate(train_iter):
            net.train()
            trainer.zero_grad()
            X, y = X.to(device), y.to(device)
            y_hat = net(X)
            l = loss(y_hat, y)
            l.backward()
            trainer.step()
            with torch.no_grad():
                metric.add(l * X.shape[0], d2l.accuracy(y_hat, y), X.shape[0])
            train_loss = metric[0] / metric[2]
            train_acc = metric[1] / metric[2]
            if (i + 1) % 50 == 0:
                animator.add(epoch + i / len(train_iter),
                             (train_loss, train_acc, None))

        test_acc = d2l.evaluate_accuracy_gpu(net, test_iter)
        animator.add(epoch+1, (None, None, test_acc))

        if scheduler:
            if scheduler.__module__ == lr_scheduler.__name__:
                # Using PyTorch In-Built scheduler
                scheduler.step()
            else:
                # Using custom defined scheduler
                for param_group in trainer.param_groups:
                    param_group['lr'] = scheduler(epoch)

    print(f'train loss {train_loss:.3f}, train acc {train_acc:.3f}, '
          f'test acc {test_acc:.3f}')
```

Let's have a look at what happens if we invoke this algorithm with default settings, such as a learning rate of 0.3 and train for 30 iterations. Note how the training accuracy keeps on increasing while progress in terms of test accuracy stalls beyond a point. The gap between both curves indicates overfitting.

```
lr, num_epochs = 0.3, 30
net = net_fn()
trainer = torch.optim.SGD(net.parameters(), lr=lr)
train(net, train_iter, test_iter, num_epochs, loss, trainer, device)
```

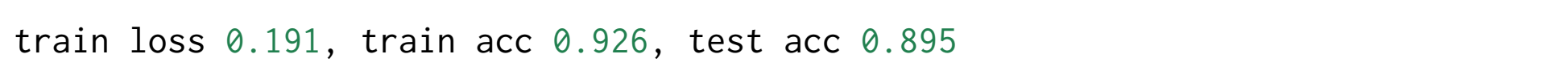

```
train loss 0.191, train acc 0.926, test acc 0.895
```

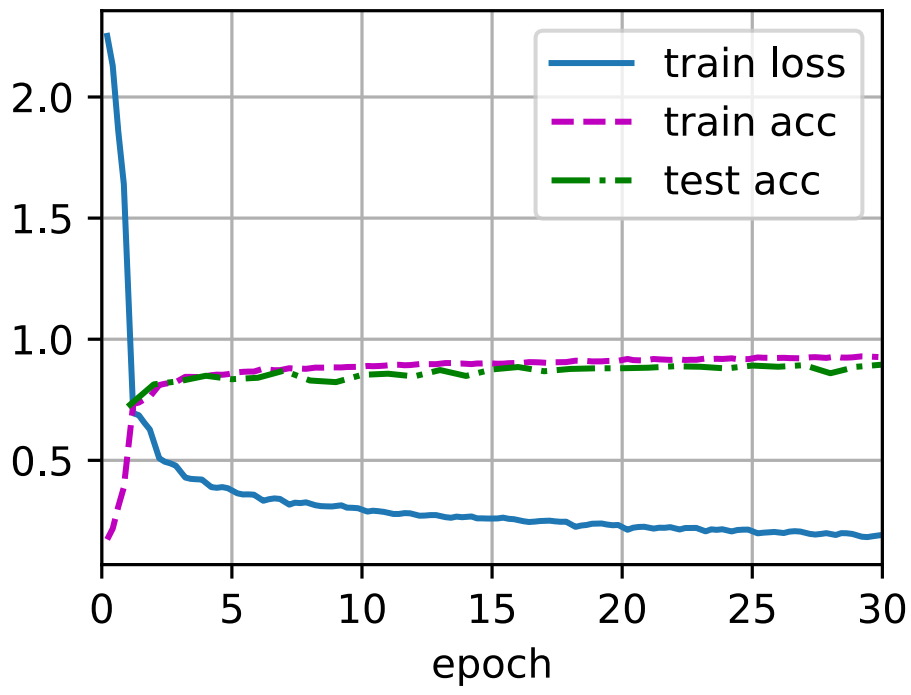

## 12.11.2 Schedulers

One way of adjusting the learning rate is to set it explicitly at each step. This is conveniently achieved by the `set_learning_rate` method. We could adjust it downward after every epoch (or even after every minibatch), e.g., in a dynamic manner in response to how optimization is progressing.

```
lr = 0.1
trainer.param_groups[0]["lr"] = lr
print(f'learning rate is now {trainer.param_groups[0]["lr"]:.2f}')
```

```
learning rate is now 0.10
```

More generally we want to define a scheduler. When invoked with the number of updates it returns the appropriate value of the learning rate. Let's define a simple one that sets the learning rate to $\eta = \eta_0 (t+1)^{-\frac{1}{2}}$.

```
class SquareRootScheduler:
    def __init__(self, lr=0.1):
        self.lr = lr

    def __call__(self, num_update):
        return self.lr * pow(num_update + 1.0, -0.5)
```

Let's plot its behavior over a range of values.

```
scheduler = SquareRootScheduler(lr=0.1)
d2l.plot(torch.arange(num_epochs), [scheduler(t) for t in range(num_epochs)])
```

Now let's see how this plays out for training on Fashion-MNIST. We simply provide the scheduler as an additional argument to the training algorithm.

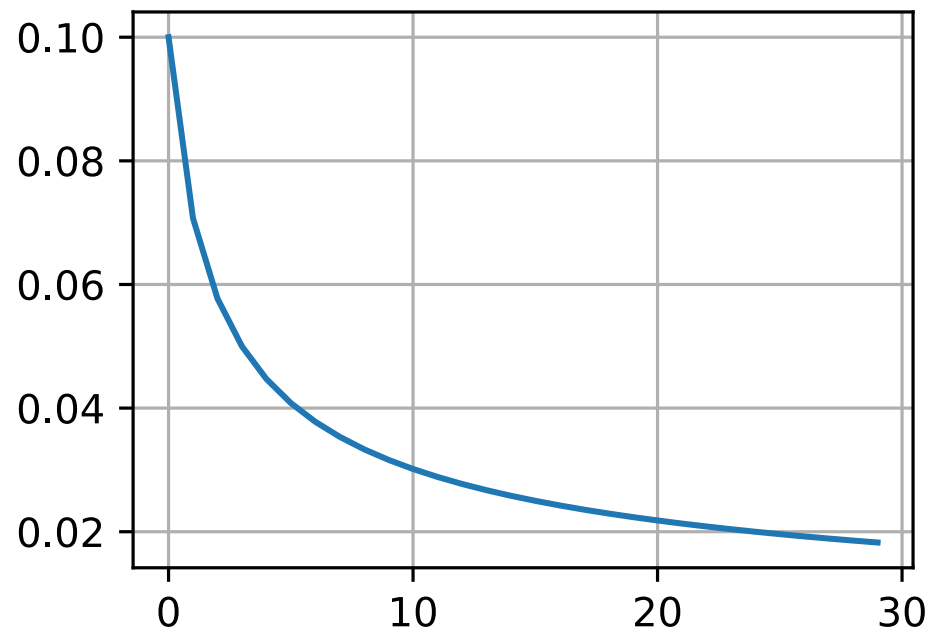

```
net = net_fn()
trainer = torch.optim.SGD(net.parameters(), lr)
train(net, train_iter, test_iter, num_epochs, loss, trainer, device,
      scheduler)
```

```
train loss 0.272, train acc 0.901, test acc 0.883
```

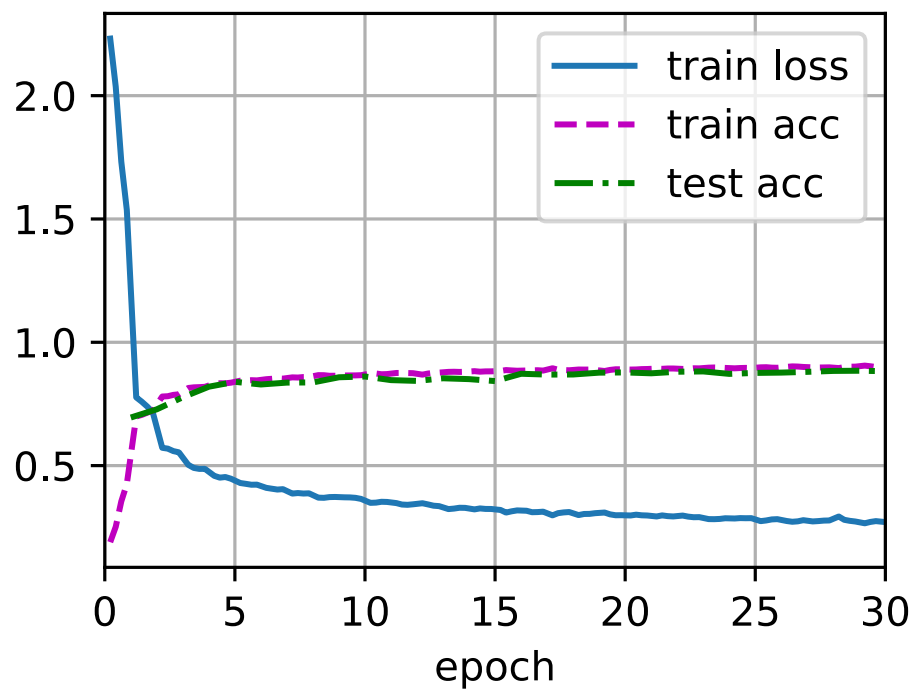

This worked quite a bit better than previously. Two things stand out: the curve was rather more smooth than previously. Secondly, there was less overfitting. Unfortunately it is not a well-resolved question as to why certain strategies lead to less overfitting in *theory*. There is some argument that a smaller stepsize will lead to parameters that are closer to zero and thus simpler. However, this does not explain the phenomenon entirely since we do not really stop early but simply reduce the learning rate gently.

### 12.11.3 Policies

While we cannot possibly cover the entire variety of learning rate schedulers, we attempt to give a brief overview of popular policies below. Common choices are polynomial decay and piecewise constant schedules. Beyond that, cosine learning rate schedules have been found to work well empirically on some problems. Lastly, on some problems it is beneficial to warm up the optimizer prior to using large learning rates.

## Factor Scheduler

One alternative to a polynomial decay would be a multiplicative one, that is $\eta_{t+1} \leftarrow \eta_t \cdot \alpha$ for $\alpha \in (0, 1)$. To prevent the learning rate from decaying beyond a reasonable lower bound the update equation is often modified to $\eta_{t+1} \leftarrow \max(\eta_{\min}, \eta_t \cdot \alpha)$.

```
class FactorScheduler:
    def __init__(self, factor=1, stop_factor_lr=1e-7, base_lr=0.1):
        self.factor = factor
        self.stop_factor_lr = stop_factor_lr
        self.base_lr = base_lr

    def __call__(self, num_update):
        self.base_lr = max(self.stop_factor_lr, self.base_lr * self.factor)
        return self.base_lr

scheduler = FactorScheduler(factor=0.9, stop_factor_lr=1e-2, base_lr=2.0)
d2l.plot(torch.arange(50), [scheduler(t) for t in range(50)])
```

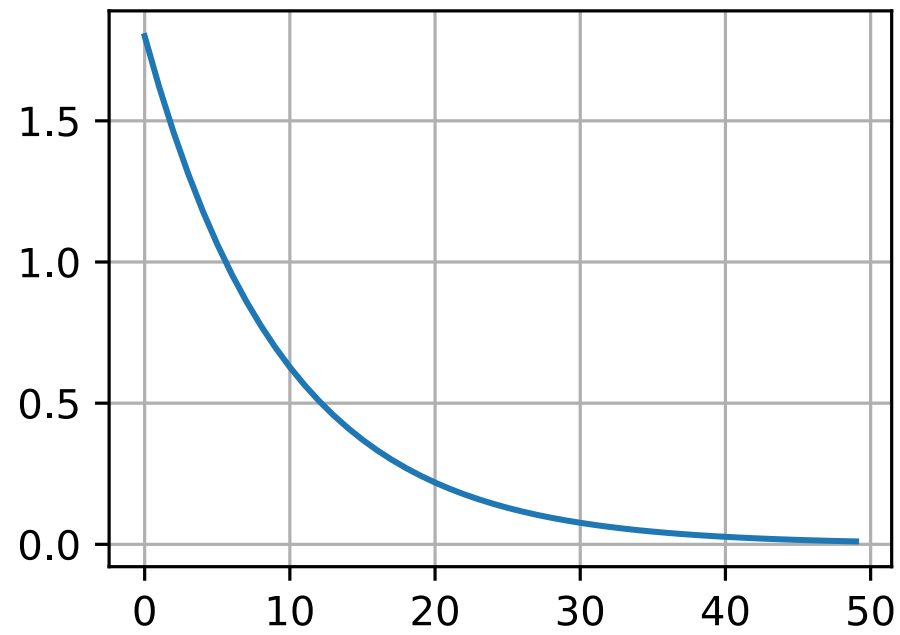

This can also be accomplished by a built-in scheduler in MXNet via the `lr_scheduler.FactorScheduler` object. It takes a few more parameters, such as warmup period, warmup mode (linear or constant), the maximum number of desired updates, etc.; Going forward we will use the built-in schedulers as appropriate and only explain their functionality here. As illustrated, it is fairly straightforward to build your own scheduler if needed.

## Multi Factor Scheduler

A common strategy for training deep networks is to keep the learning rate piecewise constant and to decrease it by a given amount every so often. That is, given a set of times when to decrease the rate, such as $s = \{5, 10, 20\}$ decrease $\eta_{t+1} \leftarrow \eta_t \cdot \alpha$ whenever $t \in s$. Assuming that the values are halved at each step we can implement this as follows.

```
net = net_fn()
trainer = torch.optim.SGD(net.parameters(), lr=0.5)
scheduler = lr_scheduler.MultiStepLR(trainer, milestones=[15, 30], gamma=0.5)

def get_lr(trainer, scheduler):
    lr = scheduler.get_last_lr()[0]
    trainer.step()
    scheduler.step()
    return lr

d2l.plot(torch.arange(num_epochs), [get_lr(trainer, scheduler)
                                  for t in range(num_epochs)])
```

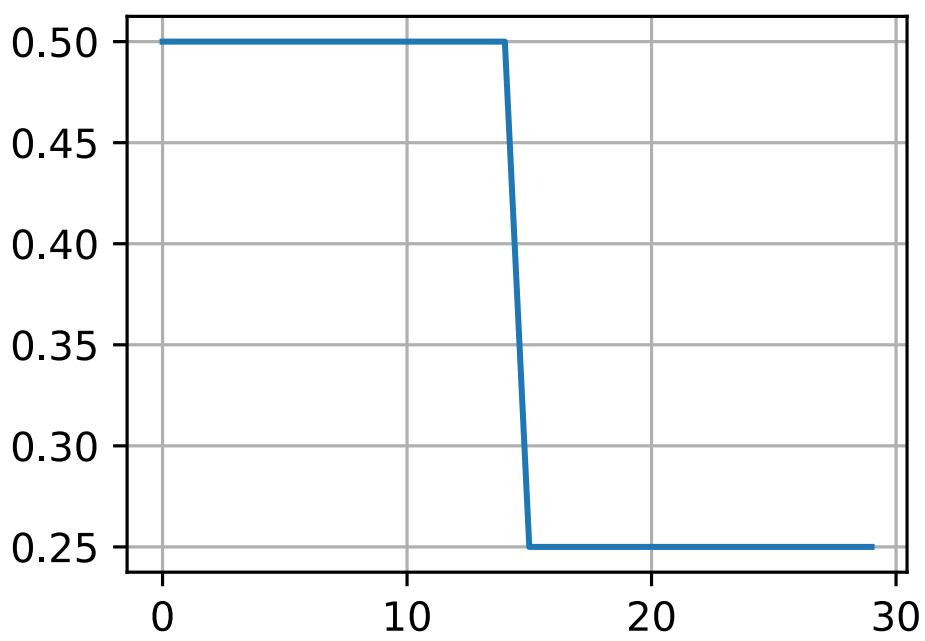

The intuition behind this piecewise constant learning rate schedule is that one lets optimization proceed until a stationary point has been reached in terms of the distribution of weight vectors. Then (and only then) do we decrease the rate such as to obtain a higher quality proxy to a good local minimum. The example below shows how this can produce ever slightly better solutions.

```
train(net, train_iter, test_iter, num_epochs, loss, trainer, device,
      scheduler)
```

```
train loss 0.196, train acc 0.926, test acc 0.889
```

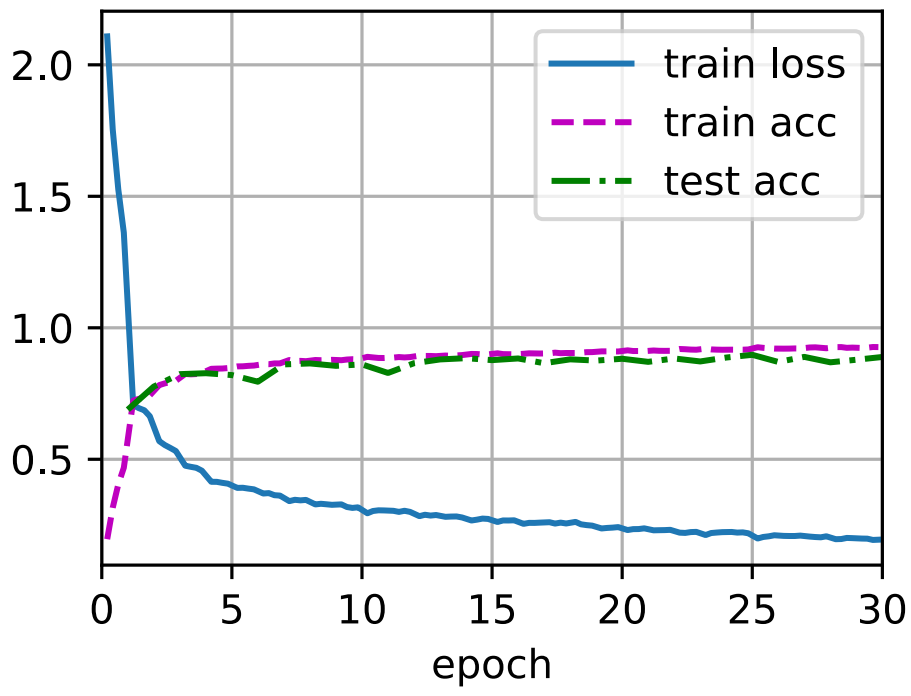

### Cosine Scheduler

A rather perplexing heuristic was proposed by Loshchilov and Hutter (2016). It relies on the observation that we might not want to decrease the learning rate too drastically in the beginning and moreover, that we might want to "refine" the solution in the end using a very small learning rate. This results in a cosine-like schedule with the following functional form for learning rates in the range $t \in [0, T]$.

$$\eta_t = \eta_T + \frac{\eta_0 - \eta_T}{2} \left(1 + \cos(\pi t/T)\right) \tag{12.11.1}$$

Here $\eta_0$ is the initial learning rate, $\eta_T$ is the target rate at time $T$. Furthermore, for $t > T$ we simply pin the value to $\eta_T$ without increasing it again. In the following example, we set the max update step $T = 20$.

```
class CosineScheduler:
    def __init__(self, max_update, base_lr=0.01, final_lr=0,
                 warmup_steps=0, warmup_begin_lr=0):
        self.base_lr_orig = base_lr
        self.max_update = max_update
        self.final_lr = final_lr
        self.warmup_steps = warmup_steps
        self.warmup_begin_lr = warmup_begin_lr
        self.max_steps = self.max_update - self.warmup_steps

    def get_warmup_lr(self, epoch):
        increase = (self.base_lr_orig - self.warmup_begin_lr) \
                       * float(epoch) / float(self.warmup_steps)
        return self.warmup_begin_lr + increase

    def __call__(self, epoch):
        if epoch < self.warmup_steps:
            return self.get_warmup_lr(epoch)
        if epoch <= self.max_update:
            self.base_lr = self.final_lr + (
                self.base_lr_orig - self.final_lr) * (1 + math.cos(
                math.pi * (epoch - self.warmup_steps) / self.max_steps)) / 2
        return self.base_lr

scheduler = CosineScheduler(max_update=20, base_lr=0.3, final_lr=0.01)
d2l.plot(torch.arange(num_epochs), [scheduler(t) for t in range(num_epochs)])
```

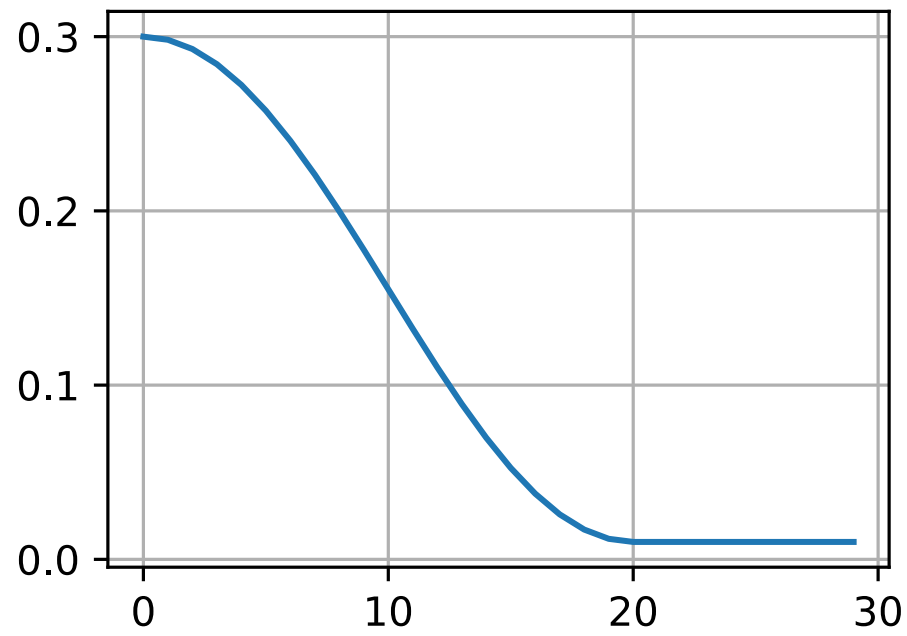

In the context of computer vision this schedule *can* lead to improved results. Note, though, that such improvements are not guaranteed (as can be seen below).

```
net = net_fn()
trainer = torch.optim.SGD(net.parameters(), lr=0.3)
train(net, train_iter, test_iter, num_epochs, loss, trainer, device,
      scheduler)
```

```
train loss 0.186, train acc 0.932, test acc 0.901
```

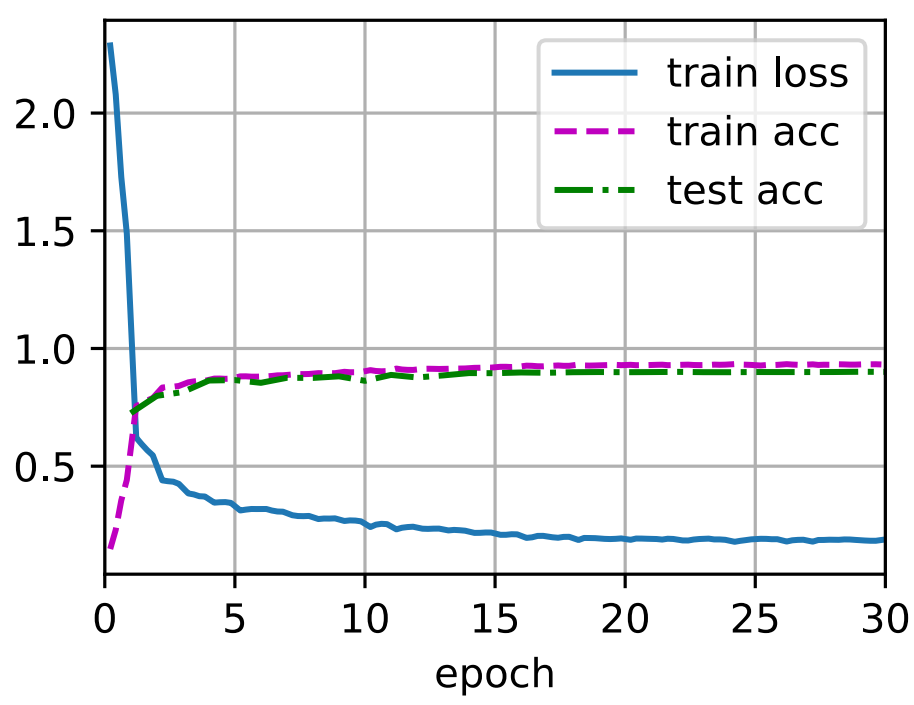

### Warmup

In some cases initializing the parameters is not sufficient to guarantee a good solution. This is particularly a problem for some advanced network designs that may lead to unstable optimization problems. We could address this by choosing a sufficiently small learning rate to prevent divergence in the beginning. Unfortunately this means that progress is slow. Conversely, a large learning rate initially leads to divergence.

A rather simple fix for this dilemma is to use a warmup period during which the learning rate *increases* to its initial maximum and to cool down the rate until the end of the optimization process. For simplicity one typically uses a linear increase for this purpose. This leads to a schedule of the form indicated below.

```
scheduler = CosineScheduler(20, warmup_steps=5, base_lr=0.3, final_lr=0.01)
d2l.plot(torch.arange(num_epochs), [scheduler(t) for t in range(num_epochs)])
```

Note that the network converges better initially (in particular observe the performance during the first 5 epochs).

```
net = net_fn()
trainer = torch.optim.SGD(net.parameters(), lr=0.3)
train(net, train_iter, test_iter, num_epochs, loss, trainer, device,
      scheduler)
```

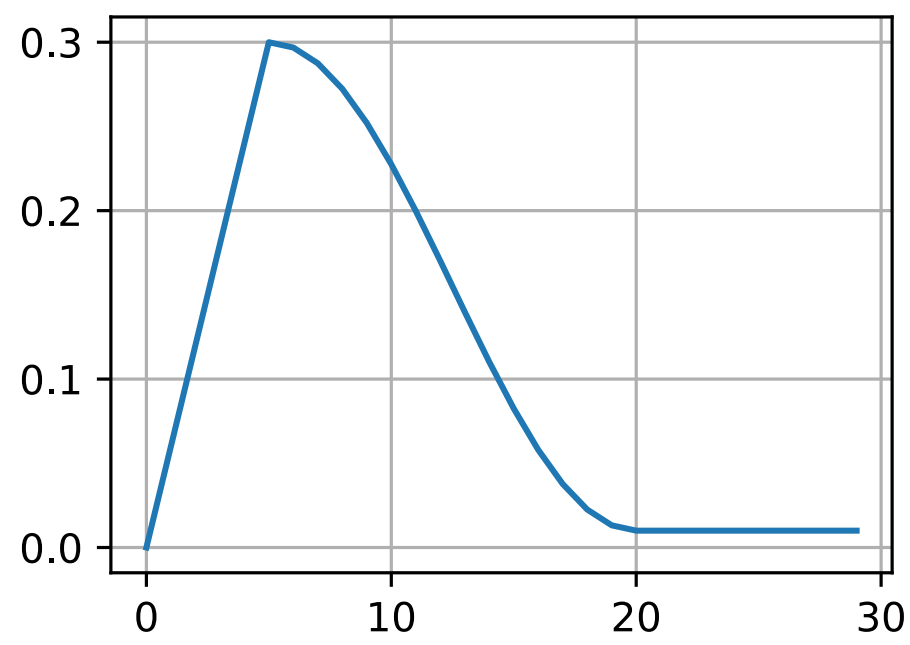

```
train loss 0.200, train acc 0.925, test acc 0.897
```

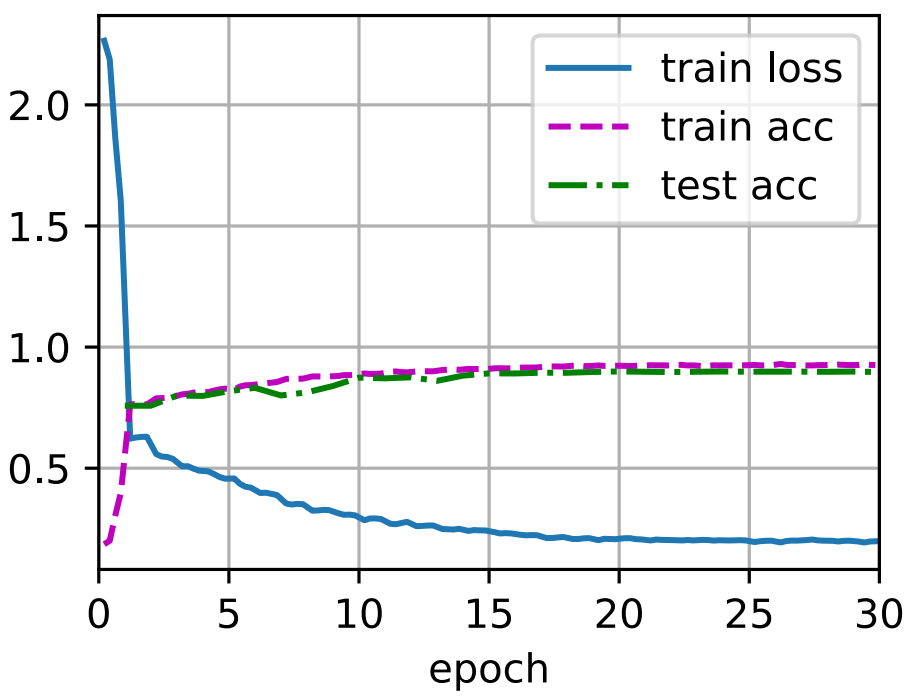

Warmup can be applied to any scheduler (not just cosine). For a more detailed discussion of learning rate schedules and many more experiments see also (Gotmare *et al.*, 2018). In particular they find that a warmup phase limits the amount of divergence of parameters in very deep networks. This makes intuitively sense since we would expect significant divergence due to random initialization in those parts of the network that take the most time to make progress in the beginning.

## 12.11.4 Summary

- Decreasing the learning rate during training can lead to improved accuracy and (most perplexingly) reduced overfitting of the model.
- A piecewise decrease of the learning rate whenever progress has plateaued is effective in practice. Essentially this ensures that we converge efficiently to a suitable solution and only then reduce the inherent variance of the parameters by reducing the learning rate.
- Cosine schedulers are popular for some computer vision problems. See e.g., GluonCV[182] for details of such a scheduler.
- A warmup period before optimization can prevent divergence.

- Optimization serves multiple purposes in deep learning. Besides minimizing the training objective, different choices of optimization algorithms and learning rate scheduling can lead to rather different amounts of generalization and overfitting on the test set (for the same amount of training error).

### 12.11.5 Exercises

1. Experiment with the optimization behavior for a given fixed learning rate. What is the best model you can obtain this way?
2. How does convergence change if you change the exponent of the decrease in the learning rate? Use `PolyScheduler` for your convenience in the experiments.
3. Apply the cosine scheduler to large computer vision problems, e.g., training ImageNet. How does it affect performance relative to other schedulers?
4. How long should warmup last?
5. Can you connect optimization and sampling? Start by using results from Welling and Teh (2011) on Stochastic Gradient Langevin Dynamics.

Discussions[183].

# 13 Computational Performance

In deep learning, datasets and models are usually large, which involves heavy computation. Therefore, computational performance matters a lot. This chapter will focus on the major factors that affect computational performance: imperative programming, symbolic programming, asynchronous computing, automatic parallelism, and multi-GPU computation. By studying this chapter, you may further improve computational performance of those models implemented in the previous chapters, for example, by reducing training time without affecting accuracy.

## 13.1 Compilers and Interpreters

So far, this book has focused on imperative programming, which makes use of statements such as `print`, `+`, and `if` to change a program's state. Consider the following example of a simple imperative program.

```
def add(a, b):
    return a + b

def fancy_func(a, b, c, d):
    e = add(a, b)
    f = add(c, d)
    g = add(e, f)
    return g

print(fancy_func(1, 2, 3, 4))
```

```
10
```

Python is an *interpreted language*. When evaluating the above `fancy_func` function it performs the operations making up the function's body *in sequence*. That is, it will evaluate `e = add(a, b)` and store the results as variable `e`, thereby changing the program's state. The next two statements `f = add(c, d)` and `g = add(e, f)` will be executed similarly, performing additions and storing the results as variables. Fig. 13.1.1 illustrates the flow of data.

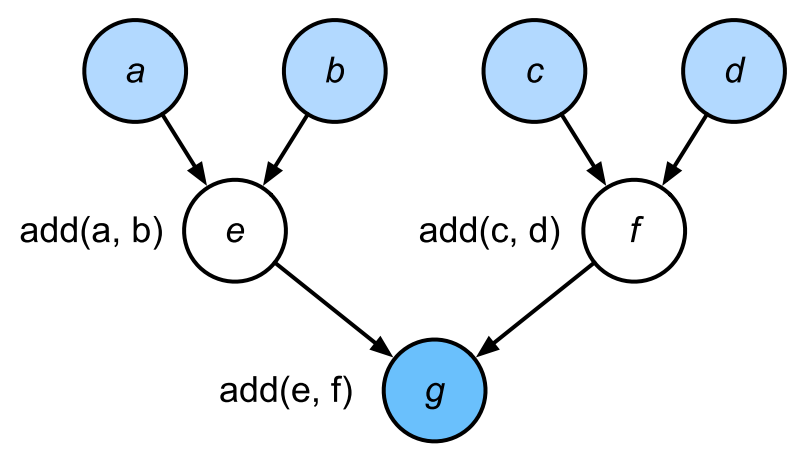

Fig. 13.1.1 Data flow in an imperative program.

Although imperative programming is convenient, it may be inefficient. On the one hand, even if the add function is repeatedly called throughout `fancy_func`, Python will execute the three function calls individually. If these are executed, say, on a GPU (or even on multiple GPUs), the overhead arising from the Python interpreter can become overwhelming. Moreover, it will need to save the variable values of `e` and `f` until all the statements in `fancy_func` have been executed. This is because we do not know whether the variables `e` and `f` will be used by other parts of the program after the statements `e = add(a, b)` and `f = add(c, d)` are executed.

## 13.1.1 Symbolic Programming

Consider the alternative, *symbolic programming*, where computation is usually performed only once the process has been fully defined. This strategy is used by multiple deep learning frameworks, including Theano and TensorFlow (the latter has acquired imperative extensions). It usually involves the following steps:

1. Define the operations to be executed.

2. Compile the operations into an executable program.

3. Provide the required inputs and call the compiled program for execution.

This allows for a significant amount of optimization. First, we can skip the Python interpreter in many cases, thus removing a performance bottleneck that can become significant on multiple fast GPUs paired with a single Python thread on a CPU. Second, a compiler might optimize and rewrite the above code into `print((1 + 2) + (3 + 4))` or even `print(10)`. This is possible since a compiler gets to see the full code before turning it into machine instructions. For instance, it can release memory (or never allocate it) whenever a variable is no longer needed. Or it can transform the code entirely into an equivalent piece. To get a better idea, consider the following simulation of imperative programming (it is Python after all) below.

```
def add_():
    return '''
def add(a, b):
    return a + b
'''
```

```
def fancy_func_():
    return '''
def fancy_func(a, b, c, d):
    e = add(a, b)
    f = add(c, d)
    g = add(e, f)
    return g
'''

def evoke_():
    return add_() + fancy_func_() + 'print(fancy_func(1, 2, 3, 4))'

prog = evoke_()
print(prog)
y = compile(prog, '', 'exec')
exec(y)
```

```
def add(a, b):
    return a + b

def fancy_func(a, b, c, d):
    e = add(a, b)
    f = add(c, d)
    g = add(e, f)
    return g
print(fancy_func(1, 2, 3, 4))
10
```

The differences between imperative (interpreted) programming and symbolic programming are as follows:

- Imperative programming is easier. When imperative programming is used in Python, the majority of the code is straightforward and easy to write. It is also easier to debug imperative programming code. This is because it is easier to obtain and print all relevant intermediate variable values, or use Python's built-in debugging tools.
- Symbolic programming is more efficient and easier to port. Symbolic programming makes it easier to optimize the code during compilation, while also having the ability to port the program into a format independent of Python. This allows the program to be run in a non-Python environment, thus avoiding any potential performance issues related to the Python interpreter.

### 13.1.2 Hybrid Programming

Historically most deep learning frameworks choose between an imperative or a symbolic approach. For example, Theano, TensorFlow (inspired by the former), Keras, and CNTK formulate models symbolically. Conversely, Chainer and PyTorch take an imperative approach. An imperative mode was added to TensorFlow 2.0 and Keras in later revisions.

As mentioned above, PyTorch is based on imperative programming and uses dynamic computation graphs. In an effort to leverage the portability and efficiency of symbolic programming, developers considered whether it would be possible to combine the benefits of both programming paradigms. This led to a torchscript that lets users develop and debug using pure imperative programming, while having the ability to convert most programs into symbolic programs to be run when product-level computing performance and deployment are required.

### 13.1.3 Hybridizing the `Sequential` Class

The easiest way to get a feel for how hybridization works is to consider deep networks with multiple layers. Conventionally the Python interpreter will need to execute the code for all layers to generate an instruction that can then be forwarded to a CPU or a GPU. For a single (fast) computing device this does not cause any major issues. On the other hand, if we use an advanced 8-GPU server such as an AWS P3dn.24xlarge instance Python will struggle to keep all GPUs busy. The single-threaded Python interpreter becomes the bottleneck here. Let's see how we can address this for significant parts of the code by replacing `Sequential` with `HybridSequential`. We begin by defining a simple MLP.

```
import torch
from torch import nn
from d2l import torch as d2l

# Factory for networks
def get_net():
    net = nn.Sequential(nn.Linear(512, 256),
            nn.ReLU(),
            nn.Linear(256, 128),
            nn.ReLU(),
            nn.Linear(128, 2))
    return net

x = torch.randn(size=(1, 512))
net = get_net()
net(x)
```

```
tensor([[-0.1197,  0.0413]], grad_fn=<AddmmBackward0>)
```

By converting the model using `torch.jit.script` function, we are able to compile and optimize the computation in the MLP. The model's computation result remains unchanged.

```
net = torch.jit.script(net)
net(x)
```

```
tensor([[-0.1197,  0.0413]], grad_fn=<AddmmBackward0>)
```

This seems almost too good to be true: write the same code as before and simply convert the model using `torch.jit.script`. Once this happens the network is optimized (we will benchmark the performance below).

## Acceleration by Hybridization

To demonstrate the performance improvement gained by compilation we compare the time needed to evaluate `net(x)` before and after hybridization. Let's define a class to measure this time first. It will come handy throughout the chapter as we set out to measure (and improve) performance.

```
#@save
class Benchmark:
    """For measuring running time."""
    def __init__(self, description='Done'):
        self.description = description

    def __enter__(self):
        self.timer = d2l.Timer()
        return self

    def __exit__(self, *args):
        print(f'{self.description}: {self.timer.stop():.4f} sec')
```

Now we can invoke the network twice, once with and once without torchscript.

```
net = get_net()
with Benchmark('Without torchscript'):
    for i in range(1000): net(x)

net = torch.jit.script(net)
with Benchmark('With torchscript'):
    for i in range(1000): net(x)
```

```
Without torchscript: 18.1045 sec
With torchscript: 20.5523 sec
```

As is observed in the above results, after an `nn.Sequential` instance is scripted using the `torch.jit.script` function, computing performance is improved through the use of symbolic programming.

## Serialization

One of the benefits of compiling the models is that we can serialize (save) the model and its parameters to disk. This allows us to store a model in a manner that is independent of the front-end language of choice. This allows us to deploy trained models to other devices and easily use other front-end programming languages. At the same time the code is often faster than what can be achieved in imperative programming. Let's see the `save` function in action.

```
net.save('my_mlp')
!ls -lh my_mlp*
```

```
-rw-rw-r-- 1 ubuntu ubuntu 652K Aug 18 22:58 my_mlp
```

### 13.1.4 Summary

- Imperative programming makes it easy to design new models since it is possible to write code with control flow and the ability to use a large amount of the Python software ecosystem.
- Symbolic programming requires that we specify the program and compile it before executing it. The benefit is improved performance.

### 13.1.5 Exercises

1. Review the models that interest you in the previous chapters. Can you improve their computational performance by reimplementing them?

Discussions[184].

## 13.2 Asynchronous Computation

Today's computers are highly parallel systems, consisting of multiple CPU cores (often multiple threads per core), multiple processing elements per GPU, and often multiple GPUs per device. In short, we can process many different things at the same time, often on different devices. Unfortunately Python is not a great way of writing parallel and asynchronous code, at least not without some extra help. After all, Python is single-threaded and this is unlikely to change in the future. Deep learning frameworks such as MXNet and TensorFlow adopt an *asynchronous programming* model to improve performance, while PyTorch uses Python's own scheduler leading to a different performance trade-off. For PyTorch, by default, GPU operations are asynchronous. When you call a function that uses the GPU, the operations are enqueued to the particular device, but not necessarily executed until later. This allows us to execute more computations in parallel, including operations on the CPU or other GPUs.

Hence, understanding how asynchronous programming works helps us to develop more efficient programs, by proactively reducing computational requirements and mutual dependencies. This allows us to reduce memory overhead and increase processor utilization.

```
import os
import subprocess
import numpy
import torch
from torch import nn
from d2l import torch as d2l
```

### 13.2.1 Asynchrony via Backend

For a warmup consider the following toy problem: we want to generate a random matrix and multiply it. Let's do that both in NumPy and in PyTorch tensor to see the difference. Note that PyTorch `tensor` is defined on a GPU.

```
# Warmup for GPU computation
device = d2l.try_gpu()
a = torch.randn(size=(1000, 1000), device=device)
b = torch.mm(a, a)

with d2l.Benchmark('numpy'):
    for _ in range(10):
        a = numpy.random.normal(size=(1000, 1000))
        b = numpy.dot(a, a)

with d2l.Benchmark('torch'):
    for _ in range(10):
        a = torch.randn(size=(1000, 1000), device=device)
        b = torch.mm(a, a)
```

```
numpy: 1.4568 sec
torch: 0.0009 sec
```

The benchmark output via PyTorch is orders of magnitude faster. NumPy dot product is executed on the CPU processor while PyTorch matrix multiplication is executed on GPU and hence the latter is expected to be much faster. But the huge time difference suggests something else must be going on. By default, GPU operations are asynchronous in PyTorch. Forcing PyTorch to finish all computation prior to returning shows what happened previously: computation is being executed by the backend while the frontend returns control to Python.

```
with d2l.Benchmark():
    for _ in range(10):
        a = torch.randn(size=(1000, 1000), device=device)
        b = torch.mm(a, a)
    torch.cuda.synchronize(device)
```

```
Done: 0.0031 sec
```

Broadly speaking, PyTorch has a frontend for direct interaction with the users, e.g., via Python, as well as a backend used by the system to perform the computation. As shown in Fig. 13.2.1, users can write PyTorch programs in various frontend languages, such as Python and C++. Regardless of the frontend programming language used, the execution of PyTorch programs occurs primarily in the backend of C++ implementations. Operations issued by the frontend language are passed on to the backend for execution. The backend manages its own threads that continuously collect and execute queued tasks. Note that for this to work the backend must be able to keep track of the dependencies between various steps in the computational graph. Hence, it is not possible to parallelize operations that depend on each other.

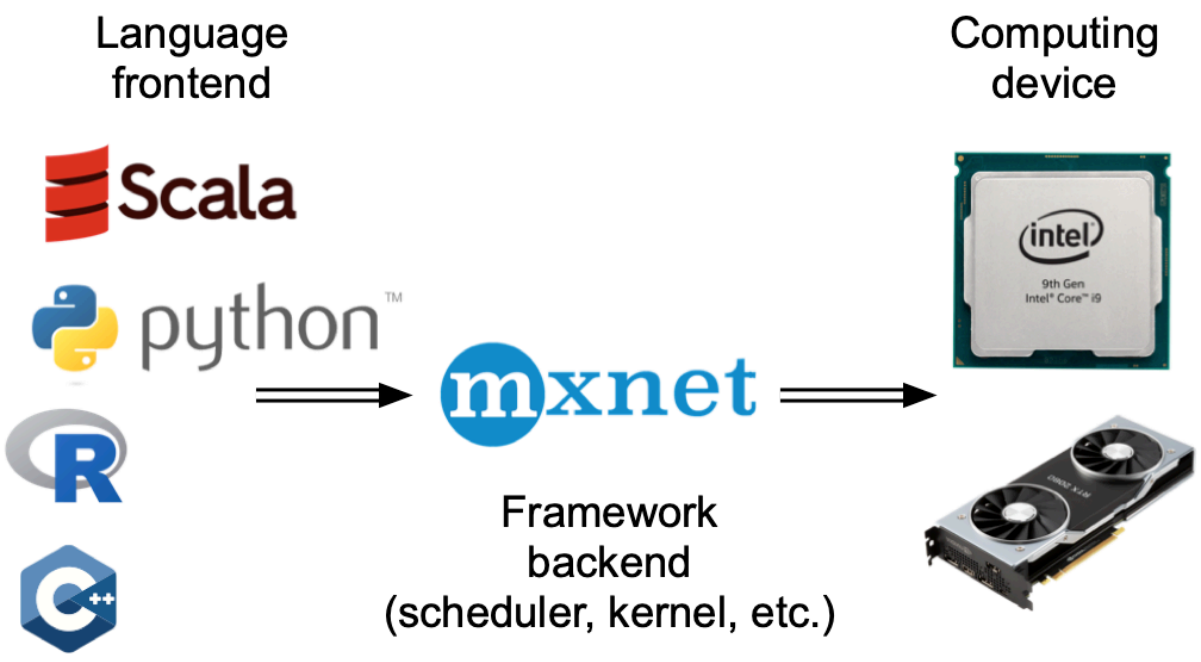

Fig. 13.2.1 Programming language frontends and deep learning framework backends.

Let's look at another toy example to understand the dependency graph a bit better.

```
x = torch.ones((1, 2), device=device)
y = torch.ones((1, 2), device=device)
z = x * y + 2
z
```

```
tensor([[3., 3.]], device='cuda:0')
```

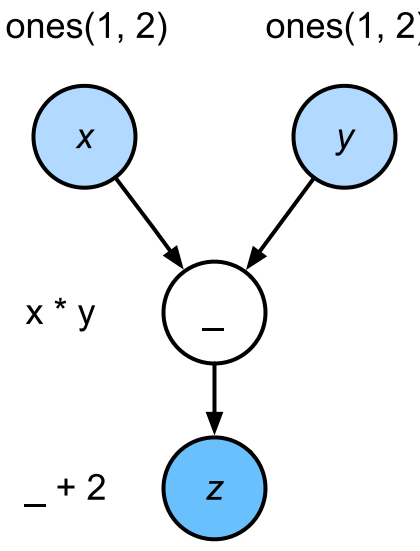

Fig. 13.2.2 The backend tracks dependencies between various steps in the computational graph.

The code snippet above is also illustrated in Fig. 13.2.2. Whenever the Python frontend thread executes one of the first three statements, it simply returns the task to the backend queue. When the last statement's results need to be *printed*, the Python frontend thread will wait for the C++ backend thread to finish computing the result of the variable z. One benefit of this design is that the Python frontend thread does not need to perform actual computations. Thus, there is little impact on the program's overall performance, regardless of Python's performance. Fig. 13.2.3 illustrates how frontend and backend interact.

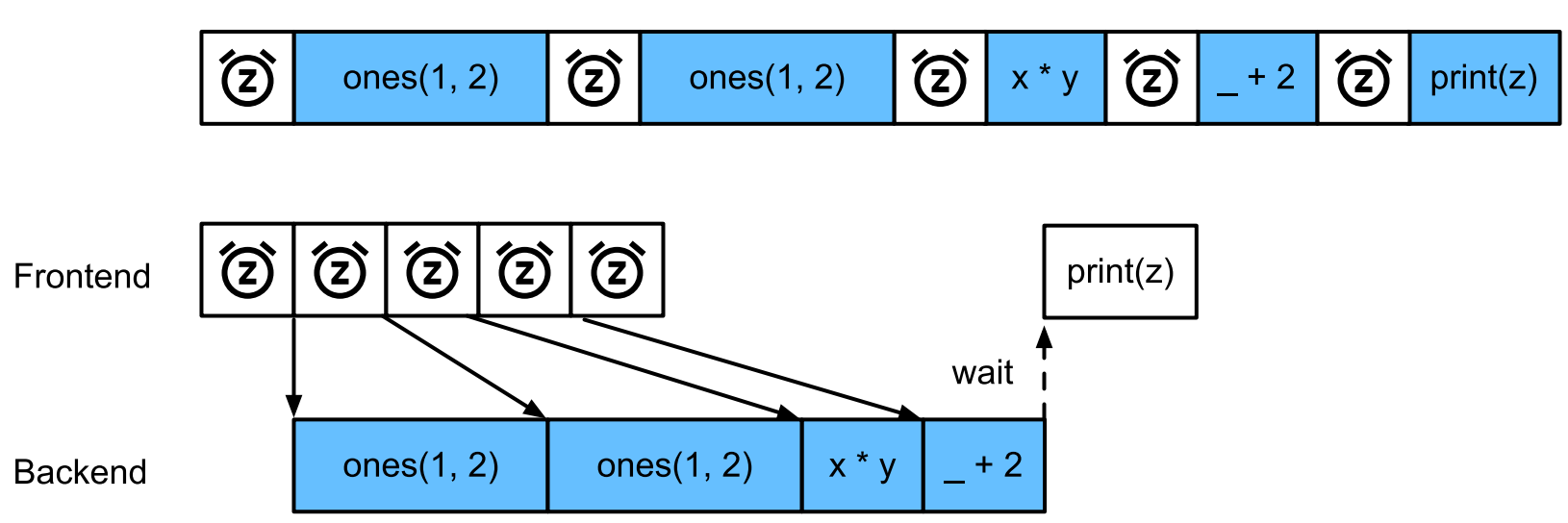

Fig. 13.2.3 Interactions of the frontend and backend.

## 13.2.2 Barriers and Blockers

## 13.2.3 Improving Computation

## 13.2.4 Summary

- Deep learning frameworks may decouple the Python frontend from an execution backend. This allows for fast asynchronous insertion of commands into the backend and associated parallelism.
- Asynchrony leads to a rather responsive frontend. However, use caution not to overfill the task queue since it may lead to excessive memory consumption. It is recommended to synchronize for each minibatch to keep frontend and backend approximately synchronized.
- Chip vendors offer sophisticated performance analysis tools to obtain a much more fine-grained insight into the efficiency of deep learning.

## 13.2.5 Exercises

1. On the CPU, benchmark the same matrix multiplication operations in this section. Can you still observe asynchrony via the backend?

Discussions[185].

# 13.3 Automatic Parallelism

Deep learning frameworks (e.g., MXNet and PyTorch) automatically construct computational graphs at the backend. Using a computational graph, the system is aware of all the dependencies, and can selectively execute multiple non-interdependent tasks in parallel to improve speed. For instance, Fig. 13.2.2 in Section 13.2 initializes two variables independently. Consequently the system can choose to execute them in parallel.

Typically, a single operator will use all the computational resources on all CPUs or on a single GPU. For example, the `dot` operator will use all cores (and threads) on all CPUs, even if there are multiple CPU processors on a single machine. The same applies to a single GPU. Hence parallelization is not quite so useful for single-device computers. With multiple devices things matter more. While parallelization is typically most relevant between multiple GPUs, adding the local CPU will increase performance slightly. For example, see Hadjis *et al.* (2016) that focuses on training computer vision models combining a GPU and a CPU. With the convenience of an automatically parallelizing framework we can accomplish the same goal in a few lines of Python code. More broadly, our discussion of automatic parallel computation focuses on parallel computation using both CPUs and GPUs, as well as the parallelization of computation and communication.

Note that we need at least two GPUs to run the experiments in this section.

```
import torch
from d2l import torch as d2l
```

## 13.3.1 Parallel Computation on GPUs

Let's start by defining a reference workload to test: the `run` function below performs 10 matrix-matrix multiplications on the device of our choice using data allocated into two variables: `x_gpu1` and `x_gpu2`.

```
devices = d2l.try_all_gpus()
def run(x):
    return [x.mm(x) for _ in range(50)]

x_gpu1 = torch.rand(size=(4000, 4000), device=devices[0])
x_gpu2 = torch.rand(size=(4000, 4000), device=devices[1])
```

Now we apply the function to the data. To ensure that caching does not play a role in the results we warm up the devices by performing a single pass on either of them prior to measuring. `torch.cuda.synchronize()` waits for all kernels in all streams on a CUDA device to complete. It takes in a `device` argument, the device for which we need to synchronize. It uses the current device, given by `current_device()`, if the device argument is `None` (default).

```
run(x_gpu1)
run(x_gpu2)  # Warm-up all devices
torch.cuda.synchronize(devices[0])
torch.cuda.synchronize(devices[1])

with d2l.Benchmark('GPU1 time'):
    run(x_gpu1)
    torch.cuda.synchronize(devices[0])

with d2l.Benchmark('GPU2 time'):
    run(x_gpu2)
    torch.cuda.synchronize(devices[1])
```

```
GPU1 time: 0.4628 sec
GPU2 time: 0.4522 sec
```

If we remove the `synchronize` statement between both tasks the system is free to parallelize computation on both devices automatically.

```
with d2l.Benchmark('GPU1 & GPU2'):
    run(x_gpu1)
    run(x_gpu2)
    torch.cuda.synchronize()
```

```
GPU1 & GPU2: 0.4630 sec
```

In the above case the total execution time is less than the sum of its parts, since the deep learning framework automatically schedules computation on both GPU devices without the need for sophisticated code on behalf of the user.

### 13.3.2 Parallel Computation and Communication

In many cases we need to move data between different devices, say between the CPU and GPU, or between different GPUs. For instance, this occurs when we want to perform distributed optimization where we need to aggregate the gradients over multiple accelerator cards. Let's simulate this by computing on the GPU and then copying the results back to the CPU.

```
def copy_to_cpu(x, non_blocking=False):
    return [y.to('cpu', non_blocking=non_blocking) for y in x]

with d2l.Benchmark('Run on GPU1'):
    y = run(x_gpu1)
    torch.cuda.synchronize()

with d2l.Benchmark('Copy to CPU'):
    y_cpu = copy_to_cpu(y)
    torch.cuda.synchronize()
```

```
Run on GPU1: 0.4649 sec
Copy to CPU: 2.5135 sec
```

This is somewhat inefficient. Note that we could already start copying parts of `y` to the CPU while the remainder of the list is still being computed. This situation occurs, e.g., when we compute the (backprop) gradient on a minibatch. The gradients of some of the parameters will be available earlier than that of others. Hence it works to our advantage to start using PCI-Express bus bandwidth while the GPU is still running. In PyTorch, several functions such as `to()` and `copy_()` admit an explicit `non_blocking` argument, which lets the caller bypass synchronization when it is unnecessary. Setting `non_blocking=True` allows us to simulate this scenario.

```
with d2l.Benchmark('Run on GPU1 and copy to CPU'):
    y = run(x_gpu1)
    y_cpu = copy_to_cpu(y, True)
    torch.cuda.synchronize()
```

```
Run on GPU1 and copy to CPU: 2.1902 sec
```

The total time required for both operations is (as expected) less than the sum of their parts. Note that this task is different from parallel computation as it uses a different resource: the bus between the CPU and GPUs. In fact, we could compute on both devices and communicate, all at the same time. As noted above, there is a dependency between computation and communication: `y[i]` must be computed before it can be copied to the CPU. Fortunately, the system can copy `y[i-1]` while computing `y[i]` to reduce the total running time.

We conclude with an illustration of the computational graph and its dependencies for a simple two-layer MLP when training on a CPU and two GPUs, as depicted in Fig. 13.3.1. It would be quite painful to schedule the parallel program resulting from this manually. This is where it is advantageous to have a graph-based computing backend for optimization.

## 13.3.3 Summary

- Modern systems have a variety of devices, such as multiple GPUs and CPUs. They can be used in parallel, asynchronously.
- Modern systems also have a variety of resources for communication, such as PCI Express, storage (typically solid-state drives or via networks), and network bandwidth. They can be used in parallel for peak efficiency.
- The backend can improve performance through automatic parallel computation and communication.

## 13.3.4 Exercises

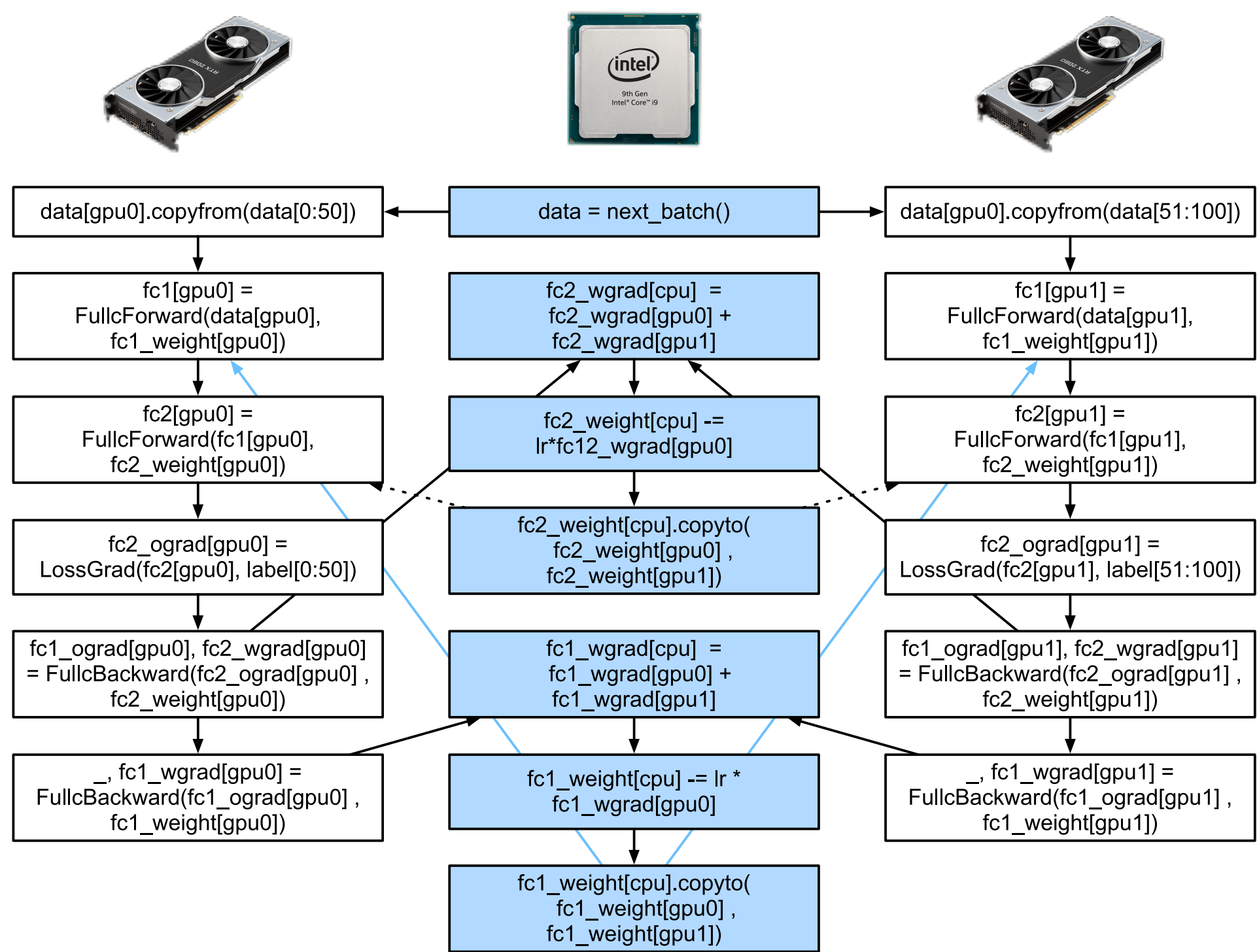

Fig. 13.3.1 The computational graph and its dependencies of a two-layer MLP on a CPU and two GPUs.

---

1. Eight operations were performed in the `run` function defined in this section. There are no dependencies between them. Design an experiment to see if the deep learning framework will automatically execute them in parallel.
2. When the workload of an individual operator is sufficiently small, parallelization can help even on a single CPU or GPU. Design an experiment to verify this.
3. Design an experiment that uses parallel computation on CPUs, GPUs, and communication between both devices.
4. Use a debugger such as NVIDIA's Nsight[186] to verify that your code is efficient.
5. Designing computation tasks that include more complex data dependencies, and run experiments to see if you can obtain the correct results while improving performance.

Discussions[187].

# 13.4 Hardware

Building systems with great performance requires a good understanding of the algorithms and models to capture the statistical aspects of the problem. At the same time it is also indispensable to have at least a modicum of knowledge of the underlying hardware. The current section is no substitute for a proper course on hardware and system design. Instead, it might serve as a starting point for understanding why some algorithms are more efficient than others and how to achieve good throughput. A good design can easily make a difference of an order of magnitude and, in turn, this can make the difference between being able to train a network (e.g., in a week) and not at all (in 3 months, thus missing the deadline). We will start by looking at computers. Then we will zoom in to look more carefully at CPUs and GPUs. Lastly we zoom out to review how multiple computers are connected in a server center or in the cloud.

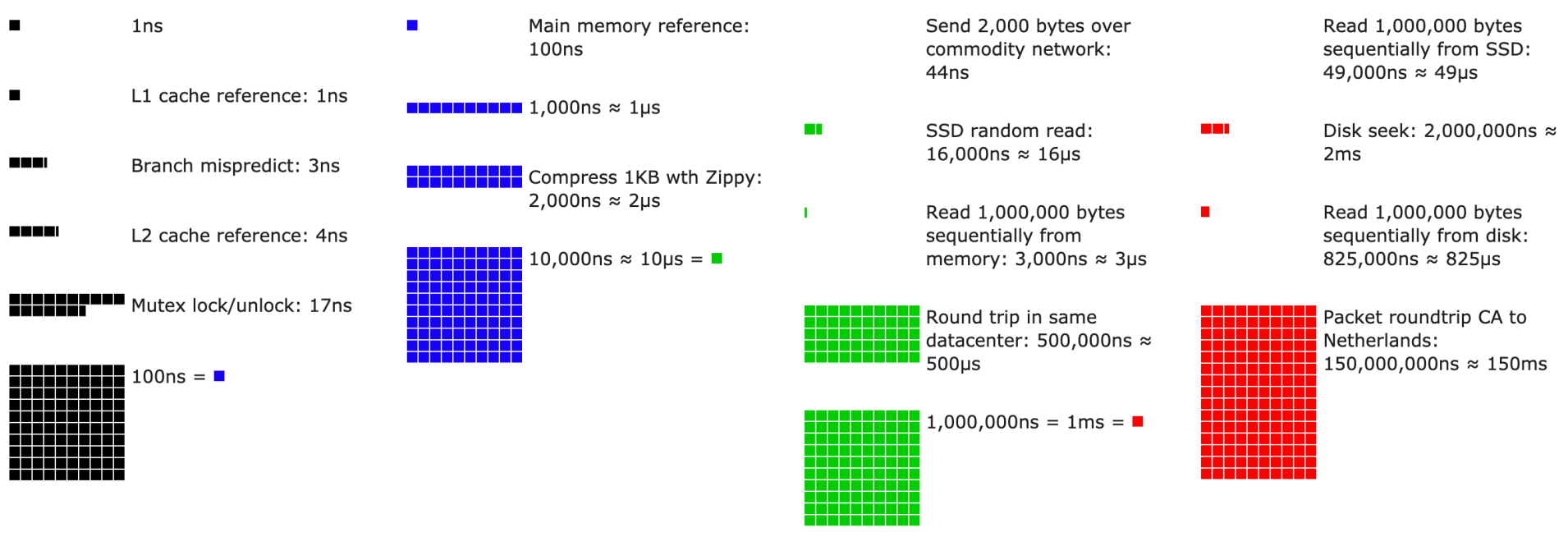

Fig. 13.4.1 Latency Numbers that every programmer should know.

Impatient readers may be able to get by with Fig. 13.4.1. It is taken from Colin Scott's interactive post[188] that gives a good overview of the progress over the past decade. The original numbers are due to Jeff Dean's Stanford talk from 2010[189]. The discussion below explains some of the rationale for these numbers and how they can guide us in designing algorithms. The discussion below is very high level and cursory. It is clearly *no substitute* for a proper course but rather just meant to provide enough information for a statistical modeler to make suitable design decisions. For an in-depth overview of computer architecture we refer the reader to (Hennessy and Patterson, 2011) or a recent course on the subject, such as the one by Arste Asanovic[190].

## 13.4.1 Computers

Most deep learning researchers and practitioners have access to a computer with a fair amount of memory, computation, some form of an accelerator such as a GPU, or multiples thereof. A computer consists of the following key components:

- A processor (also referred to as a CPU) that is able to execute the programs we give it (in

addition to running an operating system and many other things), typically consisting of 8 or more cores.

- Memory (RAM) to store and retrieve the results from computation, such as weight vectors and activations, and training data.
- An Ethernet network connection (sometimes multiple) with speeds ranging from 1 GB/s to 100 GB/s. On high end servers more advanced interconnects can be found.
- A high speed expansion bus (PCIe) to connect the system to one or more GPUs. Servers have up to 8 accelerators, often connected in an advanced topology, while desktop systems have 1 or 2, depending on the budget of the user and the size of the power supply.
- Durable storage, such as a magnetic hard disk drive, a solid state drive, in many cases connected using the PCIe bus. It provides efficient transfer of training data to the system and storage of intermediate checkpoints as needed.

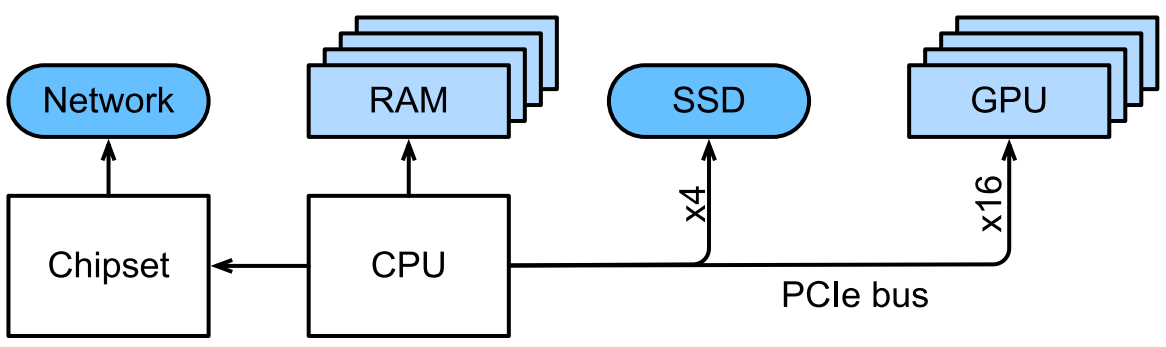

Fig. 13.4.2 Connectivity of components of a computer.

As Fig. 13.4.2 indicates, most components (network, GPU, and storage) are connected to the CPU across the PCIe bus. It consists of multiple lanes that are directly attached to the CPU. For instance AMD's Threadripper 3 has 64 PCIe 4.0 lanes, each of which is capable 16 Gbit/s data transfer in both directions. The memory is directly attached to the CPU with a total bandwidth of up to 100 GB/s.

When we run code on a computer we need to shuffle data to the processors (CPUs or GPUs), perform computation, and then move the results off the processor back to RAM and durable storage. Hence, in order to get good performance we need to make sure that this works seamlessly without any one of the systems becoming a major bottleneck. For instance, if we cannot load images quickly enough the processor will not have any work to do. Likewise, if we cannot move matrices quickly enough to the CPU (or GPU), its processing elements will starve. Finally, if we want to synchronize multiple computers across the network, the latter should not slow down computation. One option is to interleave communication and computation. Let's have a look at the various components in more detail.

## 13.4.2 Memory

At its most basic memory is used to store data that needs to be readily accessible. At present CPU RAM is typically of the DDR4[191] variety, offering 20–25 GB/s bandwidth per module. Each module has a 64-bit-wide bus. Typically pairs of memory modules are used to allow for multiple channels. CPUs have between 2 and 4 memory channels, i.e., they have between 4 0GB/s and 100 GB/s peak memory bandwidth. Often there are two banks per channel. For instance AMD's Zen 3 Threadripper has 8 slots.

While these numbers are impressive, indeed, they only tell part of the story. When we want to read a portion from memory we first need to tell the memory module where the information can be found. That is, we first need to send the *address* to RAM. Once this is accomplished we can choose to read just a single 64 bit record or a long sequence of records. The latter is called *burst read*. In a nutshell, sending an address to memory and setting up the transfer takes approximately 100 ns (details depend on the specific timing coefficients of the memory chips used), every subsequent transfer takes only 0.2 ns. In short, the first read is 500 times as expensive as subsequent ones! Note that we could perform up to 10,000,000 random reads per second. This suggests that we avoid random memory access as far as possible and use burst reads (and writes) instead.

Matters are a bit more complex when we take into account that we have multiple *banks*. Each bank can read memory largely independently. This means two things. On the one hand, the effective number of random reads is up to 4 times higher, provided that they are spread evenly across memory. It also means that it is still a bad idea to perform random reads since burst reads are 4 times faster, too. On the other hand, due to memory alignment to 64 bit boundaries it is a good idea to align any data structures with the same boundaries. Compilers do this pretty much automatically[192] when the appropriate flags are set. Curious readers are encouraged to review a lecture on DRAMs such as the one by Zeshan Chishti[193].

GPU memory is subject to even higher bandwidth requirements since they have many more processing elements than CPUs. By and large there are two options to address them. The first is to make the memory bus significantly wider. For instance, NVIDIA's RTX 2080 Ti has a 352-bit-wide bus. This allows for much more information to be transferred at the same time. Second, GPUs use specific high-performance memory. Consumer-grade devices, such as NVIDIA's RTX and Titan series typically use GDDR6[194] chips with over 500 GB/s aggregate bandwidth. An alternative is to use HBM (high bandwidth memory) modules. They use a very different interface and connect directly with GPUs on a dedicated silicon wafer. This makes them very expensive and their use is typically limited to high-end server chips, such as the NVIDIA Volta V100 series of accelerators. Quite unsurprisingly, GPU memory is generally *much* smaller than CPU memory due to the higher cost of the former. For our purposes, by and large their performance characteristics are similar, just a lot faster. We can safely ignore the details for the purpose of this book. They only matter when tuning GPU kernels for high throughput.

### 13.4.3 Storage

We saw that some of the key characteristics of RAM are *bandwidth* and *latency*. The same is true for storage devices, just that the differences can be even more extreme.

## Hard Disk Drives

*Hard disk drives* (HDDs) have been in use for over half a century. In a nutshell they contain a number of spinning platters with heads that can be positioned to read or write at any given track. High-end disks hold up to 16 TB on 9 platters. One of the key benefits of HDDs is that they are relatively inexpensive. One of their many downsides are their typically catastrophic failure modes and their relatively high read latency.

To understand the latter, consider the fact that HDDs spin at around 7,200 RPM (revolutions per minute). If they were much faster they would shatter due to the centrifugal force exerted on the platters. This has a major downside when it comes to accessing a specific sector on the disk: we need to wait until the platter has rotated in position (we can move the heads but not accelerate the actual disks). Hence it can take over 8 ms until the requested data is available. A common way this is expressed is to say that HDDs can operate at approximately 100 IOPs (input/output operations per second). This number has essentially remained unchanged for the past two decades. Worse still, it is equally difficult to increase bandwidth (it is in the order of 100–200 MB/s). After all, each head reads a track of bits, hence the bit rate only scales with the square root of the information density. As a result, HDDs are quickly becoming relegated to archival storage and low-grade storage for very large datasets.

## Solid State Drives

Solid state drives (SSDs) use flash memory to store information persistently. This allows for *much faster* access to stored records. Modern SSDs can operate at 100,000 to 500,000 IOPs, i.e., up to 3 orders of magnitude faster than HDDs. Furthermore, their bandwidth can reach 1–3GB/s, i.e., one order of magnitude faster than HDDs. These improvements sound almost too good to be true. Indeed, they come with the following caveats, due to the way SSDs are designed.

- SSDs store information in blocks (256 KB or larger). They can only be written as a whole, which takes significant time. Consequently bit-wise random writes on SSD have very poor performance. Likewise, writing data in general takes significant time since the block has to be read, erased and then rewritten with new information. By now SSD controllers and firmware have developed algorithms to mitigate this. Nonetheless, writes can be much slower, in particular for QLC (quad level cell) SSDs. The key for improved performance is to maintain a *queue* of operations, to prefer reads and to write in large blocks if possible.
- The memory cells in SSDs wear out relatively quickly (often already after a few thousand writes). Wear-level protection algorithms are able to spread the degradation over many cells. That said, it is not recommended to use SSDs for swapping files or for large aggregations of log-files.
- Lastly, the massive increase in bandwidth has forced computer designers to attach SSDs directly to the PCIe bus. The drives capable of handling this, referred to as NVMe (Non

Volatile Memory enhanced), can use up to 4 PCIe lanes. This amounts to up to 8GB/s on PCIe 4.0.

### Cloud Storage

Cloud storage provides a configurable range of performance. That is, the assignment of storage to virtual machines is dynamic, both in terms of quantity and in terms of speed, as chosen by users. We recommend that users increase the provisioned number of IOPs whenever latency is too high, e.g., during training with many small records.

## 13.4.4 CPUs

Central processing units (CPUs) are the centerpiece of any computer. They consist of a number of key components: *processor cores* that are able to execute machine code, a *bus* connecting them (the specific topology differs significantly between processor models, generations, and vendors), and *caches* to allow for higher bandwidth and lower latency memory access than what is possible by reads from main memory. Lastly, almost all modern CPUs contain *vector processing units* to aid with high performance linear algebra and convolutions, as they are common in media processing and machine learning.

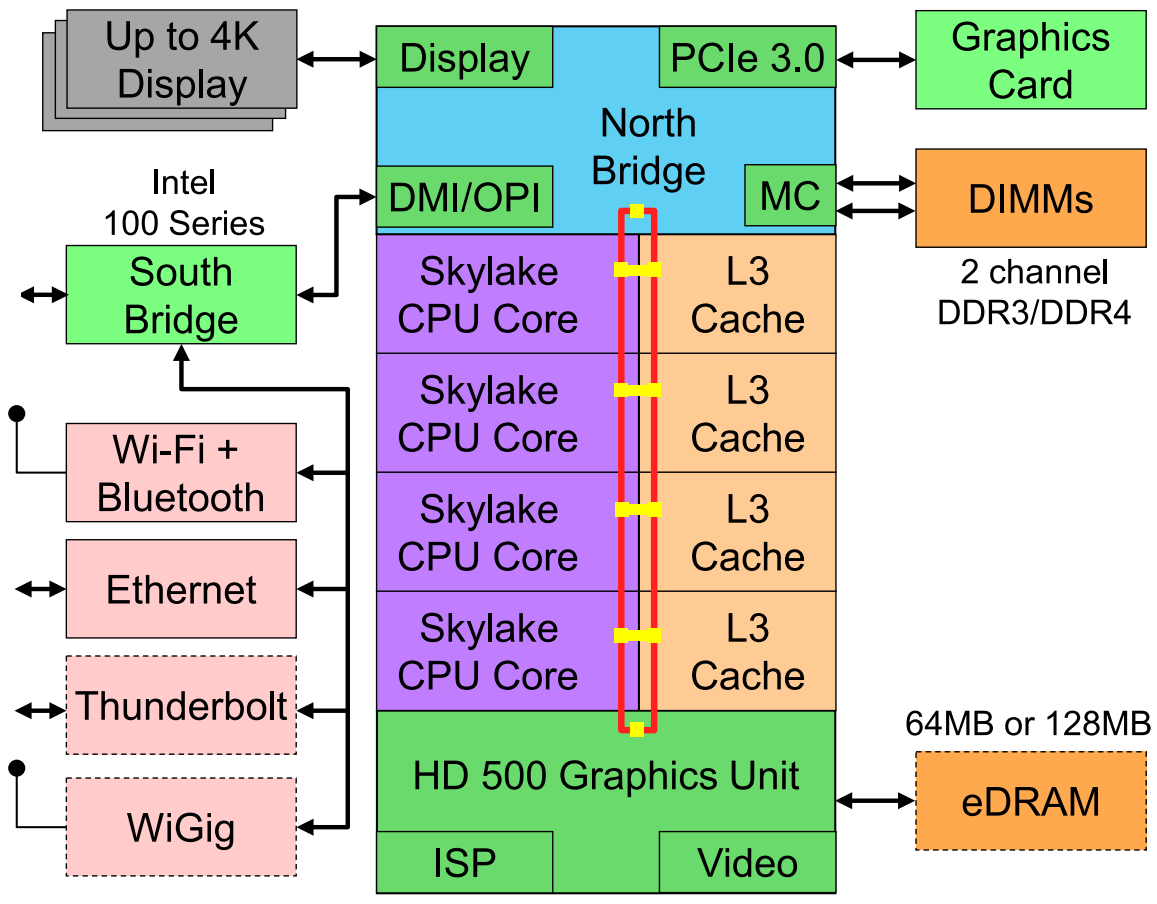

Fig. 13.4.3 Intel Skylake consumer quad-core CPU.

Fig. 13.4.3 depicts an Intel Skylake consumer-grade quad-core CPU. It has an integrated GPU, caches, and a ringbus connecting the four cores. Peripherals, such as Ethernet, WiFi, Bluetooth, SSD controller, and USB, are either part of the chipset or directly attached (PCIe) to the CPU.

### Microarchitecture

Each of the processor cores consists of a rather sophisticated set of components. While details differ between generations and vendors, the basic functionality is pretty much standard. The front-end loads instructions and tries to predict which path will be taken (e.g., for control flow). Instructions are then decoded from assembly code to microinstructions. Assembly code is often not the lowest level code that a processor executes. Instead, complex instructions may be decoded into a set of more lower level operations. These are then processed by the actual execution core. Often the latter is capable of performing many operations simultaneously. For instance, the ARM Cortex A77 core of Fig. 13.4.4 is able to perform up to 8 operations simultaneously.

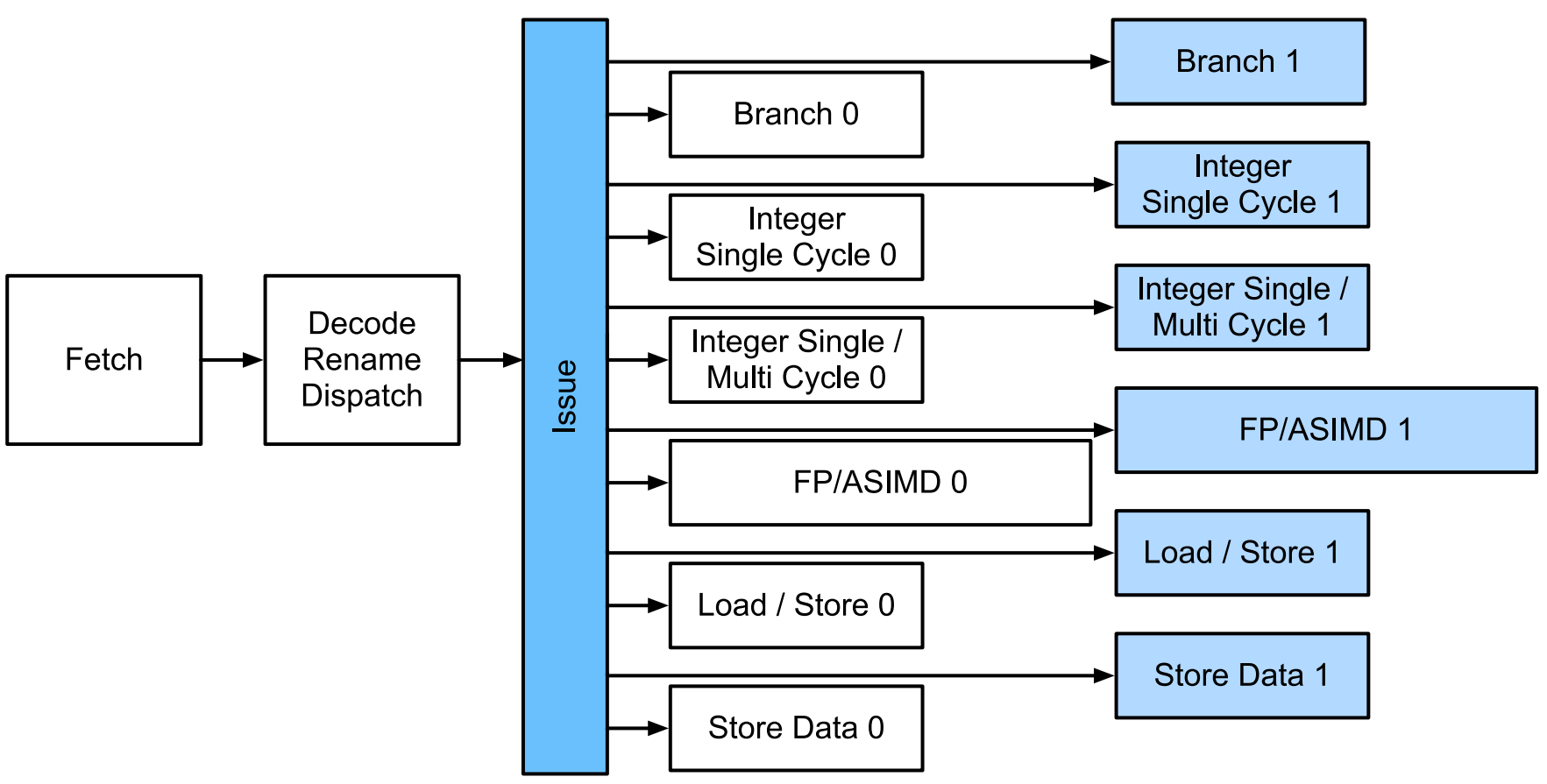

Fig. 13.4.4 ARM Cortex A77 Microarchitecture.

This means that efficient programs might be able to perform more than one instruction per clock cycle, provided that they can be carried out independently. Not all units are created equal. Some specialize in integer instructions whereas others are optimized for floating point performance. To increase throughput, the processor might also follow multiple code paths simultaneously in a branching instruction and then discard the results of the branches not taken. This is why branch prediction units matter (on the front-end) such that only the most promising paths are pursued.

### Vectorization

Deep learning is extremely compute-hungry. Hence, to make CPUs suitable for machine learning, one needs to perform many operations in one clock cycle. This is achieved via vector units. They have different names: on ARM they are called NEON, on x86 they (a recent generation) are referred to as AVX2[195] units. A common aspect is that they are able to perform SIMD (single instruction multiple data) operations. Fig. 13.4.5 shows how 8 short integers can be added in one clock cycle on ARM.

Depending on architecture choices, such registers are up to 512 bits long, allowing for the combination of up to 64 pairs of numbers. For instance, we might be multiplying two numbers and adding them to a third, which is also known as a fused multiply-add. Intel's OpenVino [196] uses these to achieve respectable throughput for deep learning on server-grade CPUs. Note, though, that this number is entirely dwarfed by what GPUs are capable of achieving. For instance, NVIDIA's RTX 2080 Ti has 4,352 CUDA cores, each of which is capable of processing such an operation at any time.

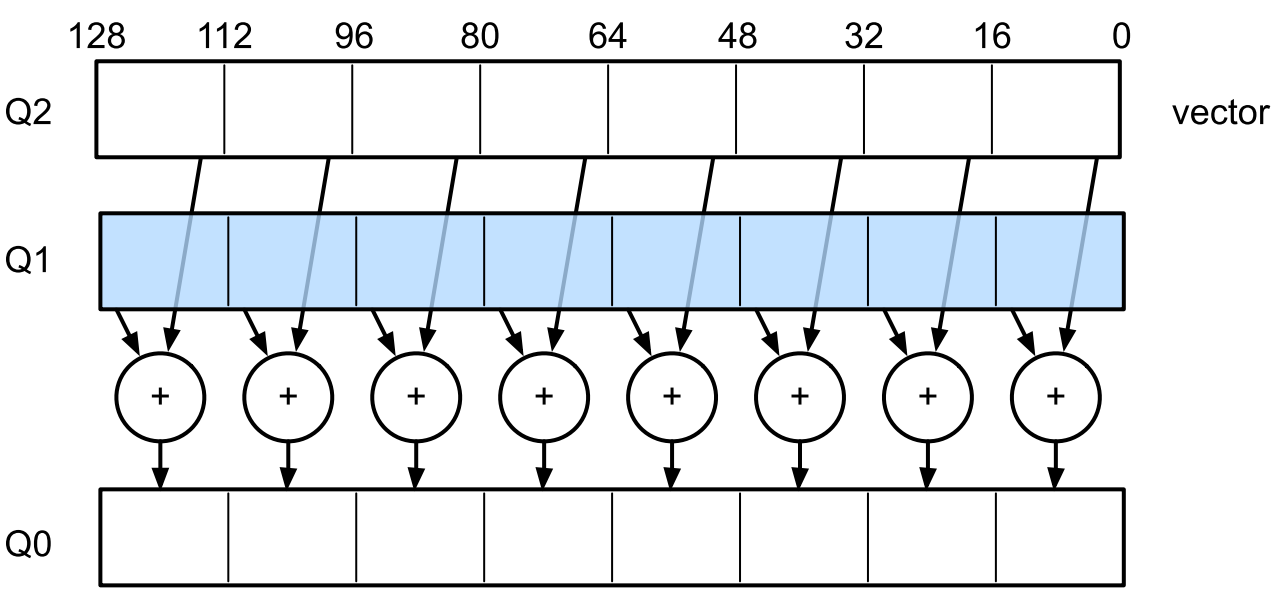

Fig. 13.4.5 128 bit NEON vectorization.

196

### Cache

Consider the following situation: we have a modest CPU core with 4 cores as depicted in Fig. 13.4.3 above, running at 2 GHz frequency. Moreover, let's assume that we have an IPC (instructions per clock) count of 1 and that the units have AVX2 with 256-bit width enabled. Let's furthermore assume that at least one of the registers used for AVX2 operations needs to be retrieved from memory. This means that the CPU consumes $4 \times 256$ bit $= 128$ bytes of data per clock cycle. Unless we are able to transfer $2 \times 10^9 \times 128 = 256 \times 10^9$ bytes to the processor per second the processing elements are going to starve. Unfortunately the memory interface of such a chip only supports 20–40 GB/s data transfer, i.e., one order of magnitude less. The fix is to avoid loading *new* data from memory as far as possible and rather to cache it locally on the CPU. This is where caches come in handy. Commonly the following names or concepts are used:

- **Registers** are strictly speaking not part of the cache. They help stage instructions. That said, CPU registers are memory locations that a CPU can access at clock speed without any delay penalty. CPUs have tens of registers. It is up to the compiler (or programmer) to use registers efficiently. For instance the C programming language has a `register` keyword.
- **L1 caches** are the first line of defense against high memory bandwidth requirements. L1 caches are tiny (typical sizes might be 32–64 KB) and often split into data and instructions caches. When data is found in the L1 cache, access is very fast. If they cannot be found there, the search progresses down the cache hierarchy.
- **L2 caches** are the next stop. Depending on architecture design and processor size they might be exclusive. They might be accessible only by a given core or shared among multiple cores. L2 caches are larger (typically 256–512 KB per core) and slower than L1. Furthermore, to access something in L2 we first need to check to realize that the data is not in L1, which adds a small amount of extra latency.

- **L3 caches** are shared among multiple cores and can be quite large. AMD's Epyc 3 server CPUs have a whopping 256 MB of cache spread across multiple chiplets. More typical numbers are in the 4–8 MB range.

Predicting which memory elements will be needed next is one of the key optimization parameters in chip design. For instance, it is advisable to traverse memory in a *forward* direction since most caching algorithms will try to *read ahead* rather than backwards. Likewise, keeping memory access patterns local is a good way of improving performance.

Adding caches is a double-edge sword. On the one hand they ensure that the processor cores do not starve of data. At the same time they increase chip size, using up area that otherwise could have been spent on increasing processing power. Moreover, *cache misses* can be expensive. Consider the worst case scenario, *false sharing*, as depicted in Fig. 13.4.6. A memory location is cached on processor 0 when a thread on processor 1 requests the data. To obtain it, processor 0 needs to stop what it is doing, write the information back to main memory and then let processor 1 read it from memory. During this operation both processors wait. Quite potentially such code runs *more slowly* on multiple processors when compared with an efficient single-processor implementation. This is one more reason for why there is a practical limit to cache sizes (besides their physical size).

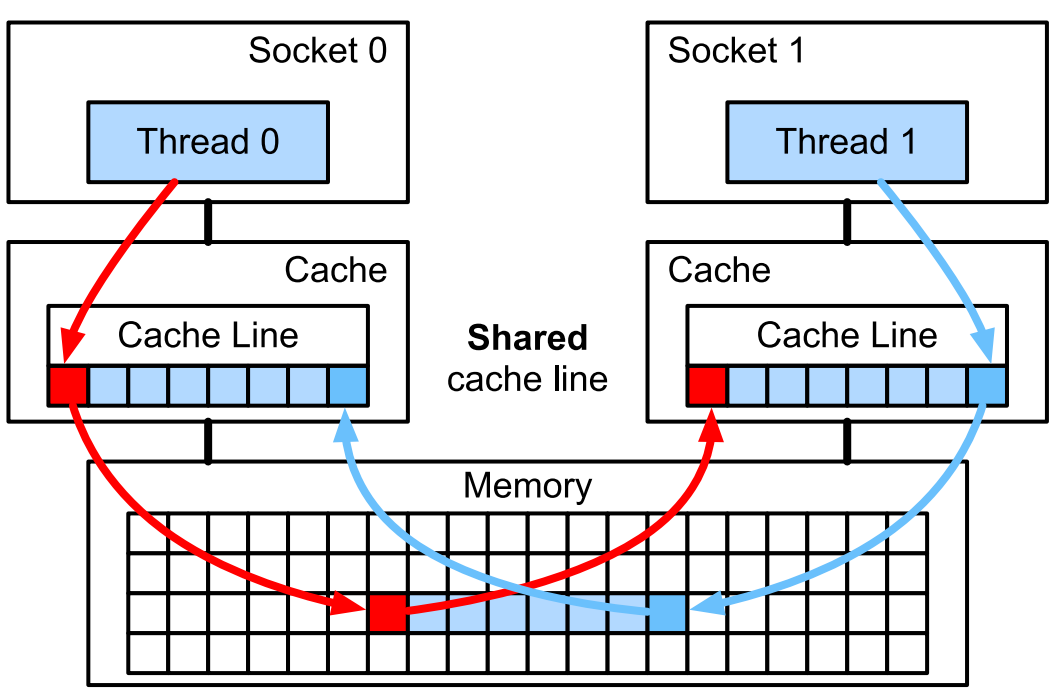

Fig. 13.4.6 False sharing (image courtesy of Intel).

## 13.4.5 GPUs and other Accelerators

It is not an exaggeration to claim that deep learning would not have been successful without GPUs. By the same token, it is quite reasonable to argue that GPU manufacturers' fortunes have increased significantly due to deep learning. This co-evolution of hardware and algorithms has led to a situation where for better or worse deep learning is the preferable statistical modeling paradigm. Hence it pays to understand the specific benefits that GPUs and related accelerators such as the TPU (Jouppi *et al.*, 2017).

Of note is a distinction that is often made in practice: accelerators are optimized either for training or inference. For the latter we only need to compute the forward propagation in a network. No storage of intermediate data is needed for backpropagation. Moreover, we may not need very precise computation (FP16 or INT8 typically suffice). On the other hand, during training all intermediate results need storage to compute gradients. Moreover, accumulating gradients requires higher precision to avoid numerical underflow (or overflow). This means that FP16 (or mixed precision with FP32) is the minimum requirement. All of this necessitates faster and larger memory (HBM2 vs. GDDR6) and more processing power. For instance, NVIDIA's Turing[197] T4 GPUs are optimized for inference whereas the V100 GPUs are preferable for training.

Recall vectorization as illustrated in Fig. 13.4.5. Adding vector units to a processor core allowed us to increase throughput significantly. For example, in the example in Fig. 13.4.5 we were able to perform 16 operations simultaneously. First, what if we added operations that optimized not just operations between vectors but also between matrices? This strategy led to tensor cores (to be covered shortly). Second, what if we added many more cores? In a nutshell, these two strategies summarize the design decisions in GPUs. Fig. 13.4.7 gives an overview of a basic processing block. It contains 16 integer and 16 floating point units. In addition to that, two tensor cores accelerate a narrow subset of additional operations relevant for deep learning. Each streaming multiprocessor consists of four such blocks.

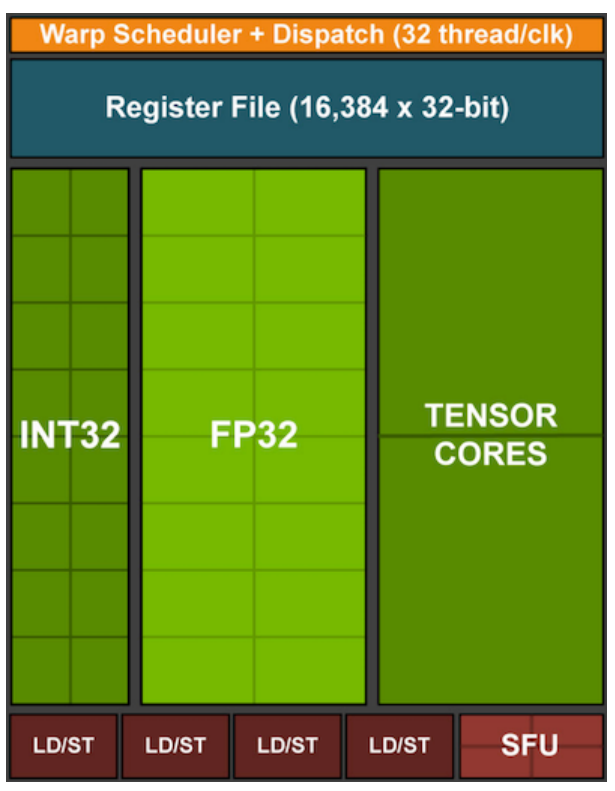

Fig. 13.4.7 NVIDIA Turing processing block (image courtesy of NVIDIA).

Next, 12 streaming multiprocessors are grouped into graphics processing clusters which make up the high-end TU102 processors. Ample memory channels and an L2 cache complement the setup. Fig. 13.4.8 has the relevant details. One of the reasons for designing such a device is that individual blocks can be added or removed as needed to allow for more compact chips and to deal with yield issues (faulty modules might not be activated). Fortunately programming such devices is well hidden from the casual deep learning researcher beneath layers of CUDA and framework code. In particular, more than one of the programs might well be executed simultaneously on the GPU, provided that there are available resources. Nonetheless it pays to be aware of the limitations of the devices to avoid picking models that do not fit into device memory.

A last aspect that is worth mentioning in more detail are *tensor cores*. They are an example of a recent trend of adding more optimized circuits that are specifically effective for deep learning. For instance, the TPU added a systolic array (Kung, 1988) for fast matrix multiplication. There the design was to support a very small number (one for the first generation of TPUs) of large operations. Tensor cores are at the other end. They are optimized for small operations involving between $4 \times 4$ and $16 \times 16$ matrices, depending on their numerical precision. Fig. 13.4.9 gives an overview of the optimizations.

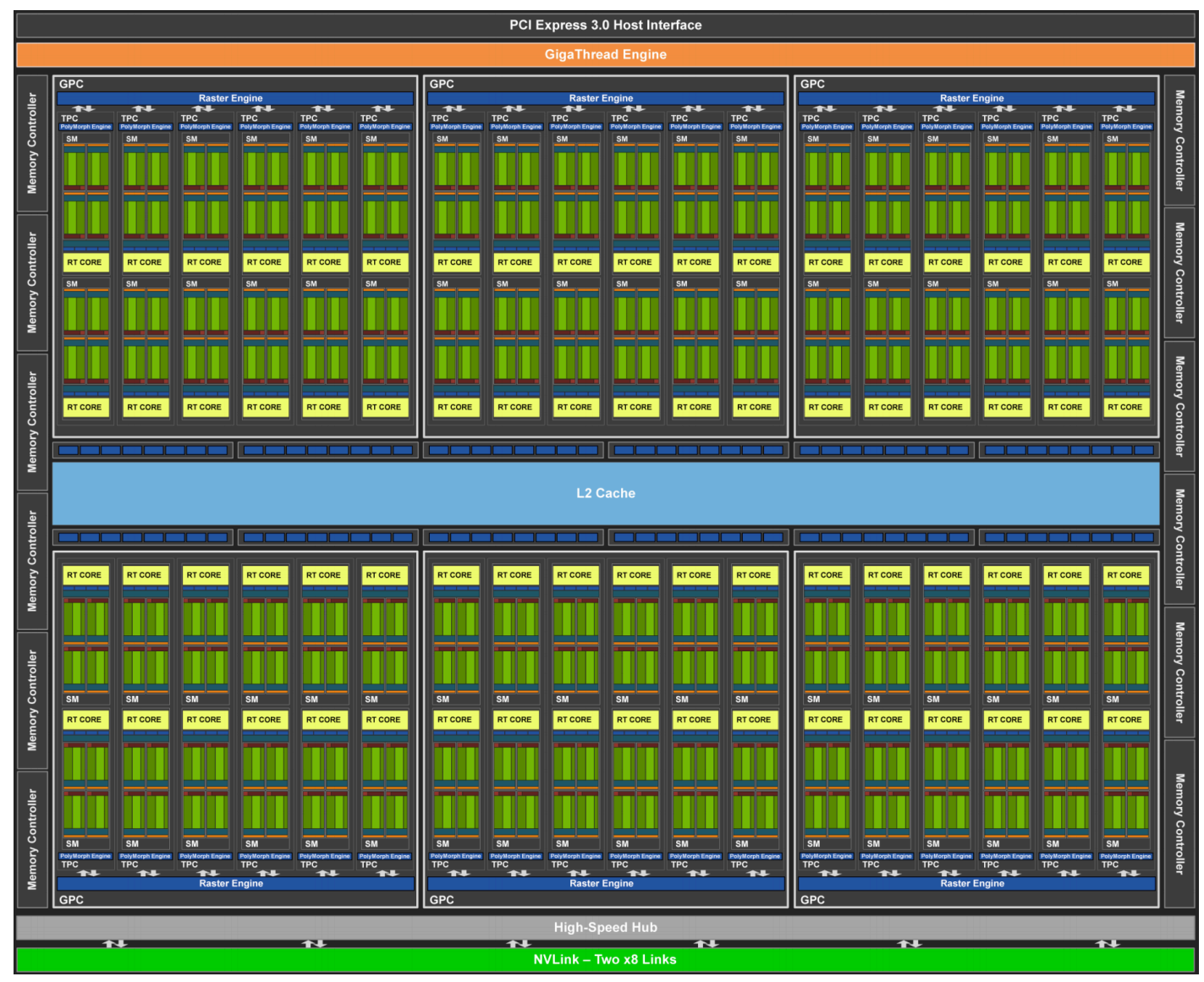

Fig. 13.4.8 NVIDIA Turing architecture (image courtesy of NVIDIA)

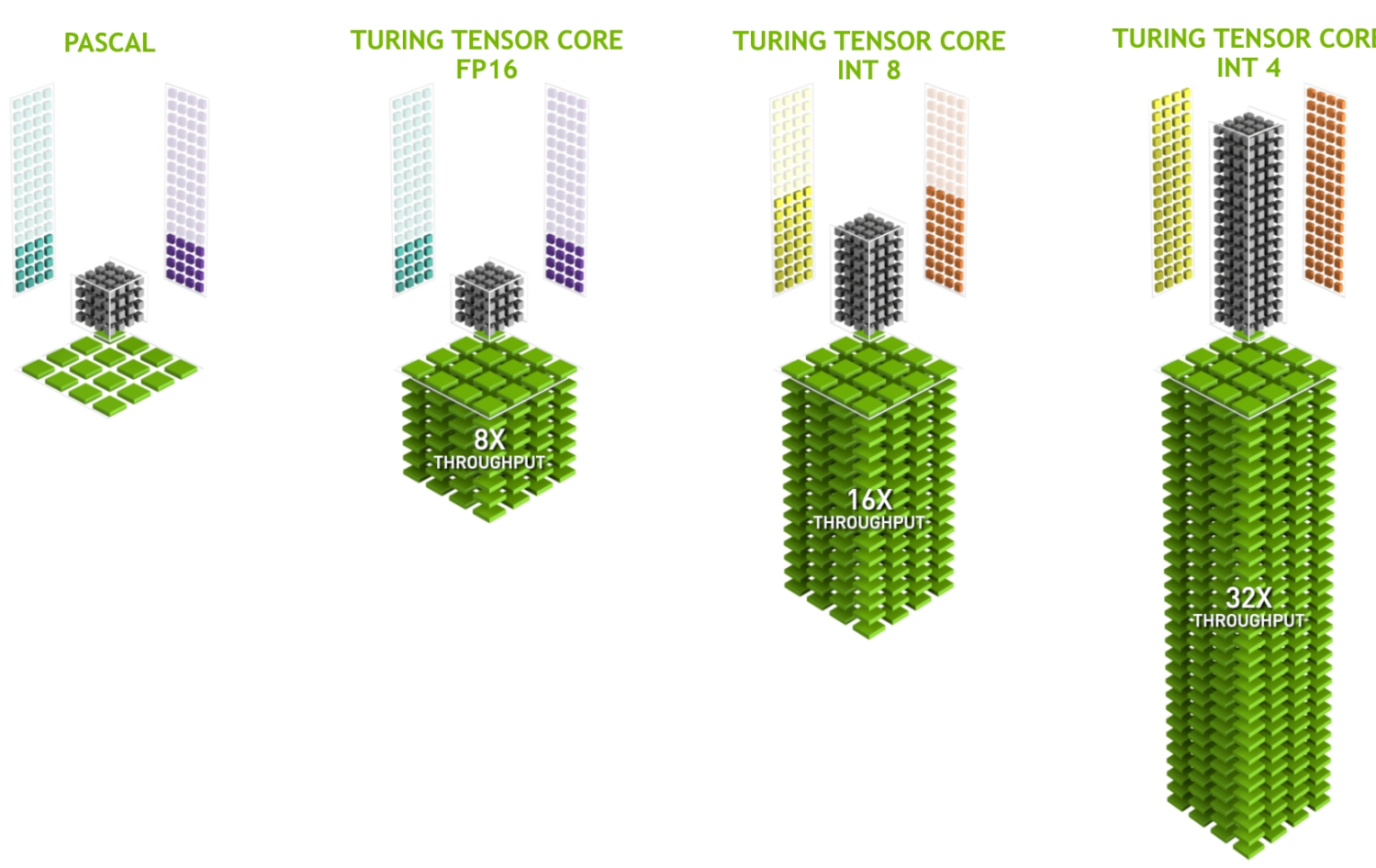

Fig. 13.4.9 NVIDIA tensor cores in Turing (image courtesy of NVIDIA).

Obviously when optimizing for computation we end up making certain compromises. One of them is that GPUs are not very good at handling interrupts and sparse data. While there

198

are notable exceptions, such as Gunrock[198] (Wang *et al.*, 2016), the access pattern of sparse matrices and vectors do not go well with the high bandwidth burst read operations where GPUs excel. Matching both goals is an area of active research. See e.g., DGL[199], a library tuned for deep learning on graphs.

199

### 13.4.6 Networks and Buses

Whenever a single device is insufficient for optimization we need to transfer data to and from it to synchronize processing. This is where networks and buses come in handy. We have a number of design parameters: bandwidth, cost, distance, and flexibility. On one end we have WiFi that has a pretty good range, is very easy to use (no wires, after all), cheap but it offers comparatively mediocre bandwidth and latency. No machine learning researcher within their right mind would use it to build a cluster of servers. In what follows we focus on interconnects that are suitable for deep learning.

- **PCIe** is a dedicated bus for very high bandwidth point-to-point connections (up to 32 GB/s on PCIe 4.0 in a 16-lane slot) per lane. Latency is in the order of single-digit microseconds (5 μs). PCIe links are precious. Processors only have a limited number of them: AMD's EPYC 3 has 128 lanes, Intel's Xeon has up to 48 lanes per chip; on desktop-grade CPUs the numbers are 20 (Ryzen 9) and 16 (Core i9) respectively. Since GPUs have typically 16 lanes, this limits the number of GPUs that can connect to the CPU at full bandwidth. After all, they need to share the links with other high bandwidth peripherals such as storage and Ethernet. Just like with RAM access, large bulk transfers are preferable due to reduced packet overhead.
- **Ethernet** is the most commonly used way of connecting computers. While it is significantly slower than PCIe, it is very cheap and resilient to install and covers much longer distances. Typical bandwidth for low-grade servers is 1 GBit/s. Higher-end devices (e.g., C5 instances[200] in the cloud) offer between 10 and 100 GBit/s bandwidth. As in all previous cases data transmission has significant overheads. Note that we almost never use raw Ethernet directly but rather a protocol that is executed on top of the physical interconnect (such as UDP or TCP/IP). This adds further overhead. Like PCIe, Ethernet is designed to connect two devices, e.g., a computer and a switch.

200

- **Switches** allow us to connect multiple devices in a manner where any pair of them can carry out a (typically full bandwidth) point-to-point connection simultaneously. For instance, Ethernet switches might connect 40 servers at high cross-sectional bandwidth. Note that switches are not unique to traditional computer networks. Even PCIe lanes can be switched[201]. This occurs, e.g., to connect a large number of GPUs to a host processor, as is the case for the P2 instances[202].

201

202

- **NVLink** is an alternative to PCIe when it comes to very high bandwidth interconnects. It offers up to 300 Gbit/s data transfer rate per link. Server GPUs (Volta V100) have six links whereas consumer-grade GPUs (RTX 2080 Ti) have only one link, operating at a reduced 100 Gbit/s rate. We recommend to use NCCL[203] to achieve high data transfer between GPUs.

203

### 13.4.7 More Latency Numbers

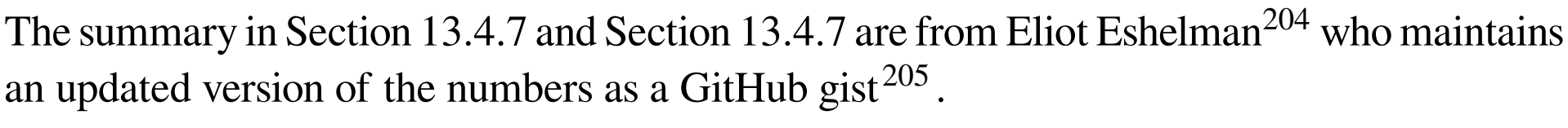

The summary in Section 13.4.7 and Section 13.4.7 are from Eliot Eshelman[204] who maintains an updated version of the numbers as a GitHub gist[205].

| Action | Time | Notes |
|---|---|---|
| L1 cache reference/hit | 1.5 ns | 4 cycles |
| Floating-point add/mult/FMA | 1.5 ns | 4 cycles |
| L2 cache reference/hit | 5 ns | 12 ~ 17 cycles |
| Branch mispredict | 6 ns | 15 ~ 20 cycles |
| L3 cache hit (unshared cache) | 16 ns | 42 cycles |
| L3 cache hit (shared in another core) | 25 ns | 65 cycles |
| Mutex lock/unlock | 25 ns | |
| L3 cache hit (modified in another core) | 29 ns | 75 cycles |
| L3 cache hit (on a remote CPU socket) | 40 ns | 100 ~ 300 cycles (40 ~ 116 ns) |
| QPI hop to a another CPU (per hop) | 40 ns | |
| 64MB memory ref. (local CPU) | 46 ns | TinyMemBench on Broadwell E5-2690v4 |
| 64MB memory ref. (remote CPU) | 70 ns | TinyMemBench on Broadwell E5-2690v4 |
| 256MB memory ref. (local CPU) | 75 ns | TinyMemBench on Broadwell E5-2690v4 |
| Intel Optane random write | 94 ns | UCSD Non-Volatile Systems Lab |
| 256MB memory ref. (remote CPU) | 120 ns | TinyMemBench on Broadwell E5-2690v4 |
| Intel Optane random read | 305 ns | UCSD Non-Volatile Systems Lab |
| Send 4KB over 100 Gbps HPC fabric | 1 μs | MVAPICH2 over Intel Omni-Path |
| Compress 1KB with Google Snappy | 3 μs | |
| Send 4KB over 10 Gbps ethernet | 10 μs | |
| Write 4KB randomly to NVMe SSD | 30 μs | DC P3608 NVMe SSD (QOS 99% is 500μs) |
| Transfer 1MB to/from NVLink GPU | 30 μs | ~33GB/s on NVIDIA 40GB NVLink |
| Transfer 1MB to/from PCI-E GPU | 80 μs | ~12GB/s on PCIe 3.0 x16 link |
| Read 4KB randomly from NVMe SSD | 120 μs | DC P3608 NVMe SSD (QOS 99%) |
| Read 1MB sequentially from NVMe SSD | 208 μs | ~4.8GB/s DC P3608 NVMe SSD |
| Write 4KB randomly to SATA SSD | 500 μs | DC S3510 SATA SSD (QOS 99.9%) |
| Read 4KB randomly from SATA SSD | 500 μs | DC S3510 SATA SSD (QOS 99.9%) |
| Round trip within same data center | 500 μs | One-way ping is ~250μs |
| Read 1MB sequentially from SATA SSD | 2 ms | ~550MB/s DC S3510 SATA SSD |
| Read 1MB sequentially from disk | 5 ms | ~200MB/s server HDD |
| Random Disk Access (seek+rotation) | 10 ms | |
| Send packet CA->Netherlands->CA | 150 ms | |

Table: Common Latency Numbers.

| Action | Time | Notes |
|---|---|---|
| GPU Shared Memory access | 30 ns | 30~90 cycles (bank conflicts add latency) |
| GPU Global Memory access | 200 ns | 200~800 cycles |
| Launch CUDA kernel on GPU | 10 μs | Host CPU instructs GPU to start kernel |
| Transfer 1MB to/from NVLink GPU | 30 μs | ~33GB/s on NVIDIA 40GB NVLink |
| Transfer 1MB to/from PCI-E GPU | 80 μs | ~12GB/s on PCI-Express x16 link |

Table: Latency Numbers for NVIDIA Tesla GPUs.

### 13.4.8 Summary

- Devices have overheads for operations. Hence it is important to aim for a small number of large transfers rather than many small ones. This applies to RAM, SSDs, networks and GPUs.
- Vectorization is key for performance. Make sure you are aware of the specific abilities of your accelerator. E.g., some Intel Xeon CPUs are particularly good for INT8 operations, NVIDIA Volta GPUs excel at FP16 matrix-matrix operations and NVIDIA Turing shines at FP16, INT8, and INT4 operations.
- Numerical overflow due to small data types can be a problem during training (and to a lesser extent during inference).
- Aliasing can significantly degrade performance. For instance, memory alignment on 64 bit CPUs should be done with respect to 64 bit boundaries. On GPUs it is a good idea to keep convolution sizes aligned, e.g., to tensor cores.
- Match your algorithms to the hardware (e.g., memory footprint, and bandwidth). Great speedup (orders of magnitude) can be achieved when fitting the parameters into caches.
- We recommend that you sketch out the performance of a novel algorithm on paper before verifying the experimental results. Discrepancies of an order-of-magnitude or more are reasons for concern.
- Use profilers to debug performance bottlenecks.
- Training and inference hardware have different sweet spots in terms of price and performance.

### 13.4.9 Exercises

1. Write C code to test whether there is any difference in speed between accessing memory aligned or misaligned relative to the external memory interface. Hint: be careful of caching effects.
2. Test the difference in speed between accessing memory in sequence or with a given stride.

3. How could you measure the cache sizes on a CPU?
4. How would you lay out data across multiple memory channels for maximum bandwidth? How would you lay it out if you had many small threads?
5. An enterprise-class HDD is spinning at 10,000 rpm. What is the absolutely minimum time an HDD needs to spend worst case before it can read data (you can assume that heads move almost instantaneously)? Why are 2.5” HDDs becoming popular for commercial servers (relative to 3.5” and 5.25” drives)?
6. Assume that an HDD manufacturer increases the storage density from 1 Tbit per square inch to 5 Tbit per square inch. How much information can you store on a ring on a 2.5” HDD? Is there a difference between the inner and outer tracks?
7. Going from 8 bit to 16 bit data types increases the amount of silicon approximately by four times. Why? Why might NVIDIA have added INT4 operations to their Turing GPUs?
8. How much faster is it to read forward through memory vs. reading backwards? Does this number differ between different computers and CPU vendors? Why? Write C code and experiment with it.
9. Can you measure the cache size of your disk? What is it for a typical HDD? Do SSDs need a cache?
10. Measure the packet overhead when sending messages across the Ethernet. Look up the difference between UDP and TCP/IP connections.
11. Direct memory access allows devices other than the CPU to write (and read) directly to (from) memory. Why is this a good idea?
12. Look at the performance numbers for the Turing T4 GPU. Why does the performance “only” double as you go from FP16 to INT8 and INT4?
13. What is the shortest time it should take for a packet on a round trip between San Francisco and Amsterdam? Hint: you can assume that the distance is 10,000 km.

Discussions[206].

## 13.5 Training on Multiple GPUs

So far we discussed how to train models efficiently on CPUs and GPUs. We even showed how deep learning frameworks allow one to parallelize computation and communication automatically between them in Section 13.3. We also showed in Section 6.7 how to list all the available GPUs on a computer using the `nvidia-smi` command. What we did *not* discuss is how to actually parallelize deep learning training. Instead, we implied in passing that one would somehow split the data across multiple devices and make it work. The present section fills in the details and shows how to train a network in parallel when starting from scratch. Details on how to take advantage of functionality in high-level APIs is relegated to Section 13.6. We assume that you are familiar with minibatch stochastic gradient descent algorithms such as the ones described in Section 12.5.

### 13.5.1 Splitting the Problem

Let's start with a simple computer vision problem and a slightly archaic network, e.g., with multiple layers of convolutions, pooling, and possibly a few fully connected layers in the end. That is, let's start with a network that looks quite similar to LeNet (LeCun *et al.*, 1998) or AlexNet (Krizhevsky *et al.*, 2012). Given multiple GPUs (2 if it is a desktop server, 4 on an AWS g4dn.12xlarge instance, 8 on a p3.16xlarge, or 16 on a p2.16xlarge), we want to partition training in a manner as to achieve good speedup while simultaneously benefitting from simple and reproducible design choices. Multiple GPUs, after all, increase both *memory* and *computation* ability. In a nutshell, we have the following choices, given a minibatch of training data that we want to classify.

First, we could partition the network across multiple GPUs. That is, each GPU takes as input the data flowing into a particular layer, processes data across a number of subsequent layers and then sends the data to the next GPU. This allows us to process data with larger networks when compared with what a single GPU could handle. Besides, memory footprint per GPU can be well controlled (it is a fraction of the total network footprint).

However, the interface between layers (and thus GPUs) requires tight synchronization. This can be tricky, in particular if the computational workloads are not properly matched between layers. The problem is exacerbated for large numbers of GPUs. The interface between layers also requires large amounts of data transfer, such as activations and gradients. This may overwhelm the bandwidth of the GPU buses. Moreover, compute-intensive, yet sequential operations are nontrivial to partition. See e.g., Mirhoseini *et al.* (2017) for a best effort in this regard. It remains a difficult problem and it is unclear whether it is possible to achieve good (linear) scaling on nontrivial problems. We do not recommend it unless there is excellent framework or operating system support for chaining together multiple GPUs.

Second, we could split the work layerwise. For instance, rather than computing 64 channels on a single GPU we could split up the problem across 4 GPUs, each of which generates data for 16 channels. Likewise, for a fully connected layer we could split the number of output units. Fig. 13.5.1 (taken from Krizhevsky *et al.* (2012)) illustrates this design, where this strategy was used to deal with GPUs that had a very small memory footprint (2 GB at the time). This allows for good scaling in terms of computation, provided that the number of channels (or units) is not too small. Besides, multiple GPUs can process increasingly larger networks since the available memory scales linearly.

However, we need a *very large* number of synchronization or barrier operations since each layer depends on the results from all the other layers. Moreover, the amount of data that needs to be transferred is potentially even larger than when distributing layers across GPUs. Thus, we do not recommend this approach due to its bandwidth cost and complexity.

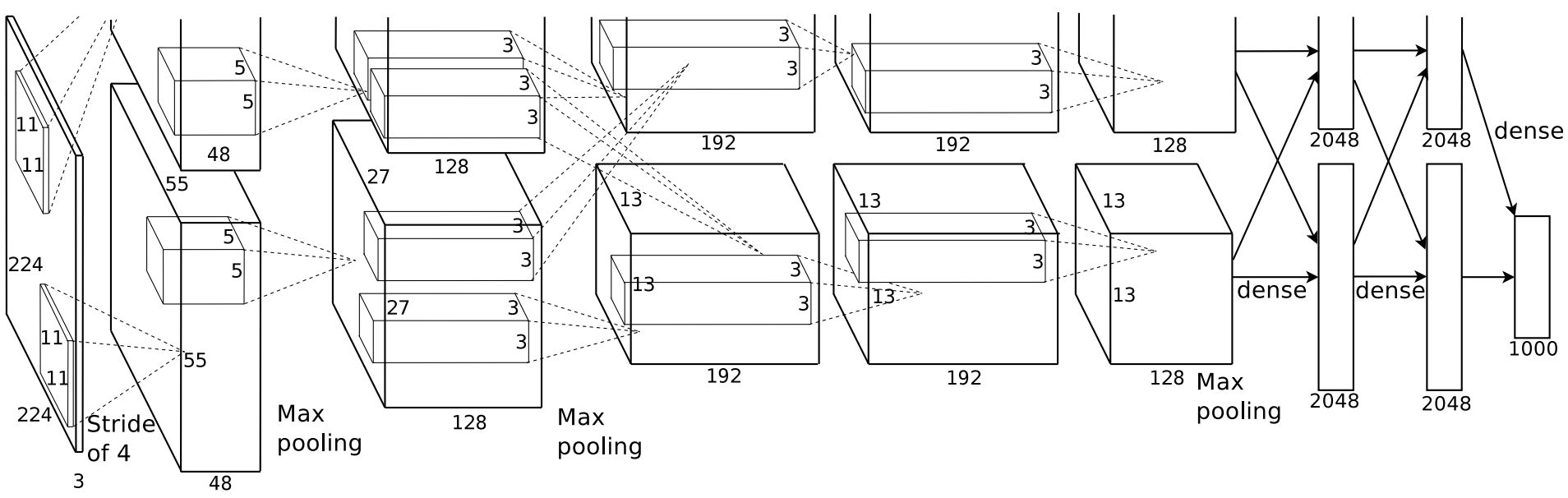

Fig. 13.5.1 Model parallelism in the original AlexNet design due to limited GPU memory.

Last, we could partition data across multiple GPUs. This way all GPUs perform the same type of work, albeit on different observations. Gradients are aggregated across GPUs after each minibatch of training data. This is the simplest approach and it can be applied in any situation. We only need to synchronize after each minibatch. That said, it is highly desirable to start exchanging gradients parameters already while others are still being computed. Moreover, larger numbers of GPUs lead to larger minibatch sizes, thus increasing training efficiency. However, adding more GPUs does not allow us to train larger models.

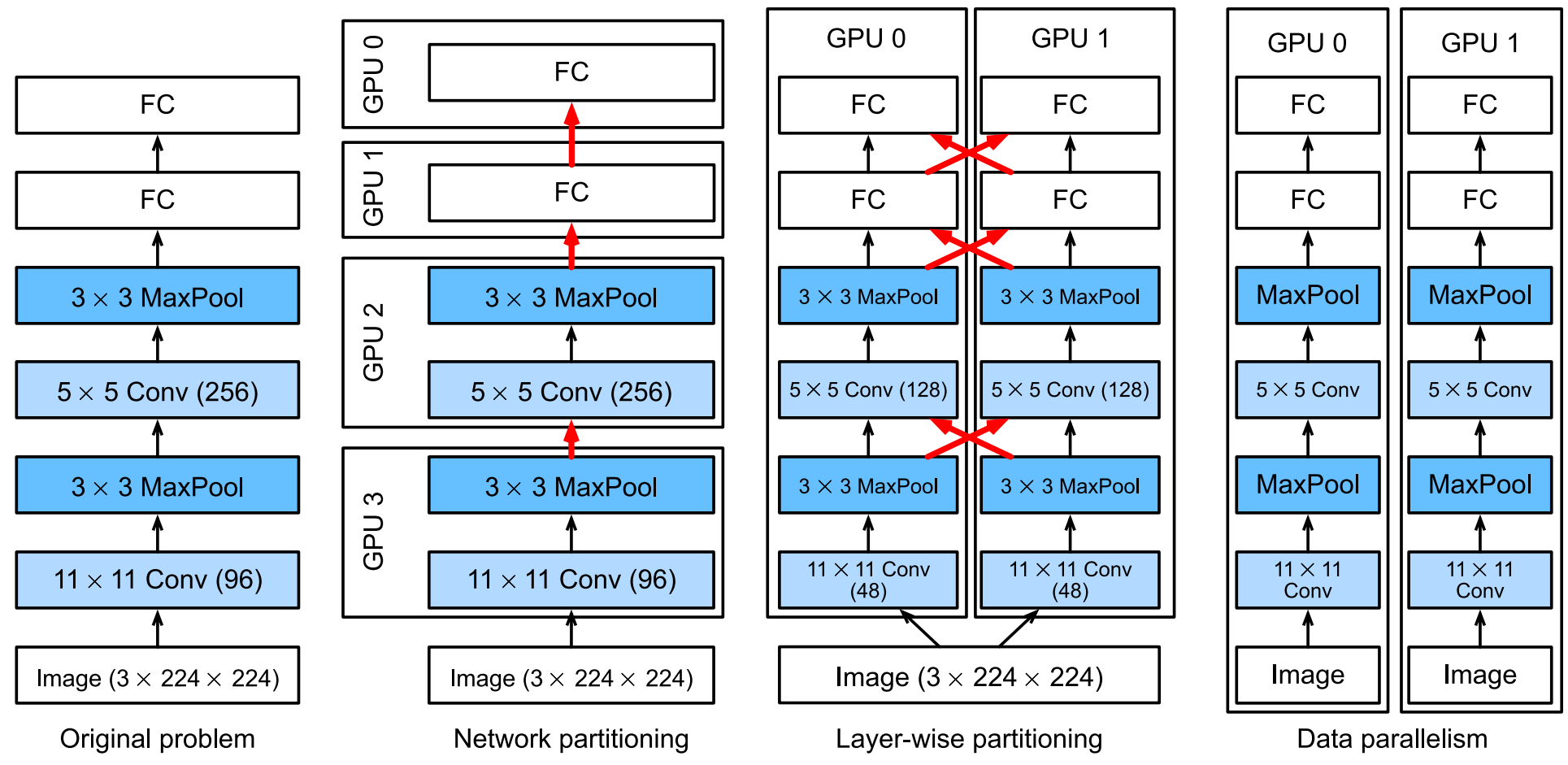

Fig. 13.5.2 Parallelization on multiple GPUs. From left to right: original problem, network partitioning, layerwise partitioning, data parallelism.

A comparison of different ways of parallelization on multiple GPUs is depicted in Fig. 13.5.2. By and large, data parallelism is the most convenient way to proceed, provided that we have access to GPUs with sufficiently large memory. See also (Li *et al.*, 2014) for a detailed description of partitioning for distributed training. GPU memory used to be a problem in the early days of deep learning. By now this issue has been resolved for all but the most unusual cases. We focus on data parallelism in what follows.

## 13.5.2 Data Parallelism

Assume that there are $k$ GPUs on a machine. Given the model to be trained, each GPU will maintain a complete set of model parameters independently though parameter values across the GPUs are identical and synchronized. As an example, Fig. 13.5.3 illustrates training with data parallelism when $k = 2$.

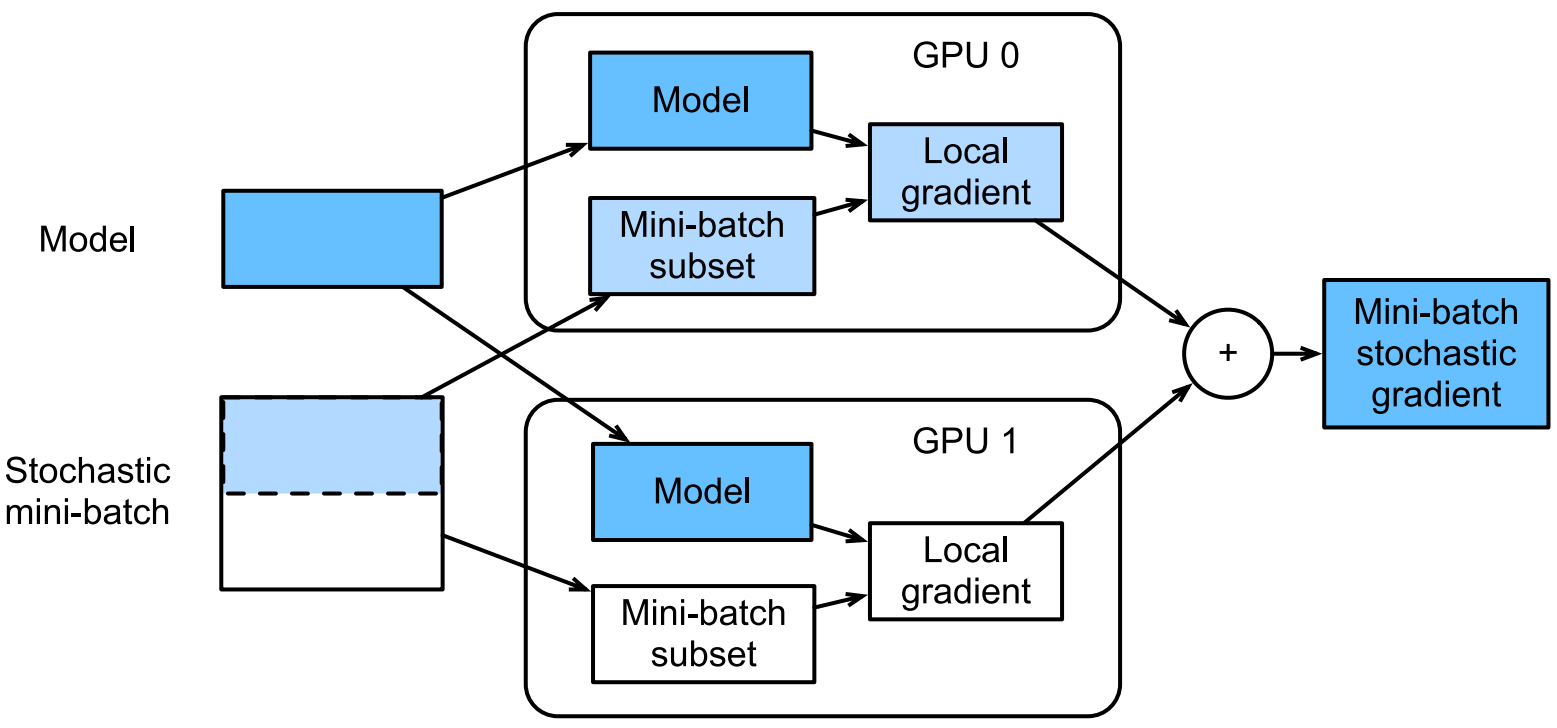

Fig. 13.5.3 Calculation of minibatch stochastic gradient descent using data parallelism on two GPUs.

In general, the training proceeds as follows:

- In any iteration of training, given a random minibatch, we split the examples in the batch into $k$ portions and distribute them evenly across the GPUs.
- Each GPU calculates loss and gradient of the model parameters based on the minibatch subset it was assigned.
- The local gradients of each of the $k$ GPUs are aggregated to obtain the current minibatch stochastic gradient.
- The aggregate gradient is re-distributed to each GPU.
- Each GPU uses this minibatch stochastic gradient to update the complete set of model parameters that it maintains.

Note that in practice we *increase* the minibatch size $k$-fold when training on $k$ GPUs such that each GPU has the same amount of work to do as if we were training on a single GPU only. On a 16-GPU server this can increase the minibatch size considerably and we may have to increase the learning rate accordingly. Also note that batch normalization in Section 8.5 needs to be adjusted, e.g., by keeping a separate batch normalization coefficient per GPU. In what follows we will use a toy network to illustrate multi-GPU training.

```
%matplotlib inline
import torch
from torch import nn
from torch.nn import functional as F
from d2l import torch as d2l
```

### 13.5.3 A Toy Network

We use LeNet as introduced in Section 7.6 (with slight modifications). We define it from scratch to illustrate parameter exchange and synchronization in detail.

```
# Initialize model parameters
scale = 0.01
W1 = torch.randn(size=(20, 1, 3, 3)) * scale
b1 = torch.zeros(20)
W2 = torch.randn(size=(50, 20, 5, 5)) * scale
b2 = torch.zeros(50)
W3 = torch.randn(size=(800, 128)) * scale
b3 = torch.zeros(128)
W4 = torch.randn(size=(128, 10)) * scale
b4 = torch.zeros(10)
params = [W1, b1, W2, b2, W3, b3, W4, b4]

# Define the model
def lenet(X, params):
    h1_conv = F.conv2d(input=X, weight=params[0], bias=params[1])
    h1_activation = F.relu(h1_conv)
    h1 = F.avg_pool2d(input=h1_activation, kernel_size=(2, 2), stride=(2, 2))
    h2_conv = F.conv2d(input=h1, weight=params[2], bias=params[3])
    h2_activation = F.relu(h2_conv)
    h2 = F.avg_pool2d(input=h2_activation, kernel_size=(2, 2), stride=(2, 2))
    h2 = h2.reshape(h2.shape[0], -1)
    h3_linear = torch.mm(h2, params[4]) + params[5]
    h3 = F.relu(h3_linear)
    y_hat = torch.mm(h3, params[6]) + params[7]
    return y_hat

# Cross-entropy loss function
loss = nn.CrossEntropyLoss(reduction='none')
```

### 13.5.4 Data Synchronization

For efficient multi-GPU training we need two basic operations. First we need to have the ability to distribute a list of parameters to multiple devices and to attach gradients (`get_params`). Without parameters it is impossible to evaluate the network on a GPU. Second, we need the ability to sum parameters across multiple devices, i.e., we need an `allreduce` function.

```
def get_params(params, device):
    new_params = [p.to(device) for p in params]
    for p in new_params:
        p.requires_grad_()
    return new_params
```

Let's try it out by copying the model parameters to one GPU.

```
new_params = get_params(params, d2l.try_gpu(0))
print('b1 weight:', new_params[1])
print('b1 grad:', new_params[1].grad)
```

```
b1 weight: tensor([0., 0., 0., 0., 0., 0., 0., 0., 0., 0., 0., 0., 0., 0., 0.,␣
↪0., 0., 0., 0., 0.],
       device='cuda:0', requires_grad=True)
b1 grad: None
```

Since we did not perform any computation yet, the gradient with regard to the bias parameter is still zero. Now let's assume that we have a vector distributed across multiple GPUs. The following `allreduce` function adds up all vectors and broadcasts the result back to all GPUs. Note that for this to work we need to copy the data to the device accumulating the results.

```
def allreduce(data):
    for i in range(1, len(data)):
        data[0][:] += data[i].to(data[0].device)
    for i in range(1, len(data)):
        data[i][:] = data[0].to(data[i].device)
```

Let's test this by creating vectors with different values on different devices and aggregate them.

```
data = [torch.ones((1, 2), device=d2l.try_gpu(i)) * (i + 1) for i in range(2)]
print('before allreduce:\n', data[0], '\n', data[1])
allreduce(data)
print('after allreduce:\n', data[0], '\n', data[1])
```

```
before allreduce:
 tensor([[1., 1.]], device='cuda:0')
 tensor([[2., 2.]], device='cuda:1')
after allreduce:
 tensor([[3., 3.]], device='cuda:0')
 tensor([[3., 3.]], device='cuda:1')
```

## 13.5.5 Distributing Data

We need a simple utility function to distribute a minibatch evenly across multiple GPUs. For instance, on two GPUs we would like to have half of the data to be copied to either of the GPUs. Since it is more convenient and more concise, we use the built-in function from the deep learning framework to try it out on a $4 \times 5$ matrix.

```
data = torch.arange(20).reshape(4, 5)
devices = [torch.device('cuda:0'), torch.device('cuda:1')]
```

```
split = nn.parallel.scatter(data, devices)
print('input :', data)
print('load into', devices)
print('output:', split)
```

```
input : tensor([[ 0,  1,  2,  3,  4],
        [ 5,  6,  7,  8,  9],
        [10, 11, 12, 13, 14],
        [15, 16, 17, 18, 19]])
load into [device(type='cuda', index=0), device(type='cuda', index=1)]
output: (tensor([[0, 1, 2, 3, 4],
        [5, 6, 7, 8, 9]], device='cuda:0'), tensor([[10, 11, 12, 13, 14],
        [15, 16, 17, 18, 19]], device='cuda:1'))
```

For later reuse we define a `split_batch` function that splits both data and labels.

```
#@save
def split_batch(X, y, devices):
    """Split `X` and `y` into multiple devices."""
    assert X.shape[0] == y.shape[0]
    return (nn.parallel.scatter(X, devices),
            nn.parallel.scatter(y, devices))
```

### 13.5.6 Training

Now we can implement multi-GPU training on a single minibatch. Its implementation is primarily based on the data parallelism approach described in this section. We will use the auxiliary functions we just discussed, `allreduce` and `split_and_load`, to synchronize the data among multiple GPUs. Note that we do not need to write any specific code to achieve parallelism. Since the computational graph does not have any dependencies across devices within a minibatch, it is executed in parallel *automatically*.

```
def train_batch(X, y, device_params, devices, lr):
    X_shards, y_shards = split_batch(X, y, devices)
    # Loss is calculated separately on each GPU
    ls = [loss(lenet(X_shard, device_W), y_shard).sum()
          for X_shard, y_shard, device_W in zip(
              X_shards, y_shards, device_params)]
    for l in ls:  # Backpropagation is performed separately on each GPU
        l.backward()
    # Sum all gradients from each GPU and broadcast them to all GPUs
    with torch.no_grad():
        for i in range(len(device_params[0])):
            allreduce([device_params[c][i].grad for c in range(len(devices))])
    # The model parameters are updated separately on each GPU
    for param in device_params:
        d2l.sgd(param, lr, X.shape[0]) # Here, we use a full-size batch
```

Now, we can define the training function. It is slightly different from the ones used in the previous chapters: we need to allocate the GPUs and copy all the model parameters to all the devices. Obviously each batch is processed using the `train_batch` function to deal with multiple GPUs. For convenience (and conciseness of code) we compute the accuracy on a single GPU, though this is *inefficient* since the other GPUs are idle.

```
def train(num_gpus, batch_size, lr):
    train_iter, test_iter = d2l.load_data_fashion_mnist(batch_size)
    devices = [d2l.try_gpu(i) for i in range(num_gpus)]
    # Copy model parameters to `num_gpus` GPUs
    device_params = [get_params(params, d) for d in devices]
    num_epochs = 10
    animator = d2l.Animator('epoch', 'test acc', xlim=[1, num_epochs])
    timer = d2l.Timer()
    for epoch in range(num_epochs):
        timer.start()
        for X, y in train_iter:
            # Perform multi-GPU training for a single minibatch
            train_batch(X, y, device_params, devices, lr)
            torch.cuda.synchronize()
        timer.stop()
        # Evaluate the model on GPU 0
        animator.add(epoch + 1, (d2l.evaluate_accuracy_gpu(
            lambda x: lenet(x, device_params[0]), test_iter, devices[0]),))
    print(f'test acc: {animator.Y[0][-1]:.2f}, {timer.avg():.1f} sec/epoch '
          f'on {str(devices)}')
```

Let's see how well this works on a single GPU. We first use a batch size of 256 and a learning rate of 0.2.

```
train(num_gpus=1, batch_size=256, lr=0.2)
```

```
test acc: 0.83, 4.1 sec/epoch on [device(type='cuda', index=0)]
```

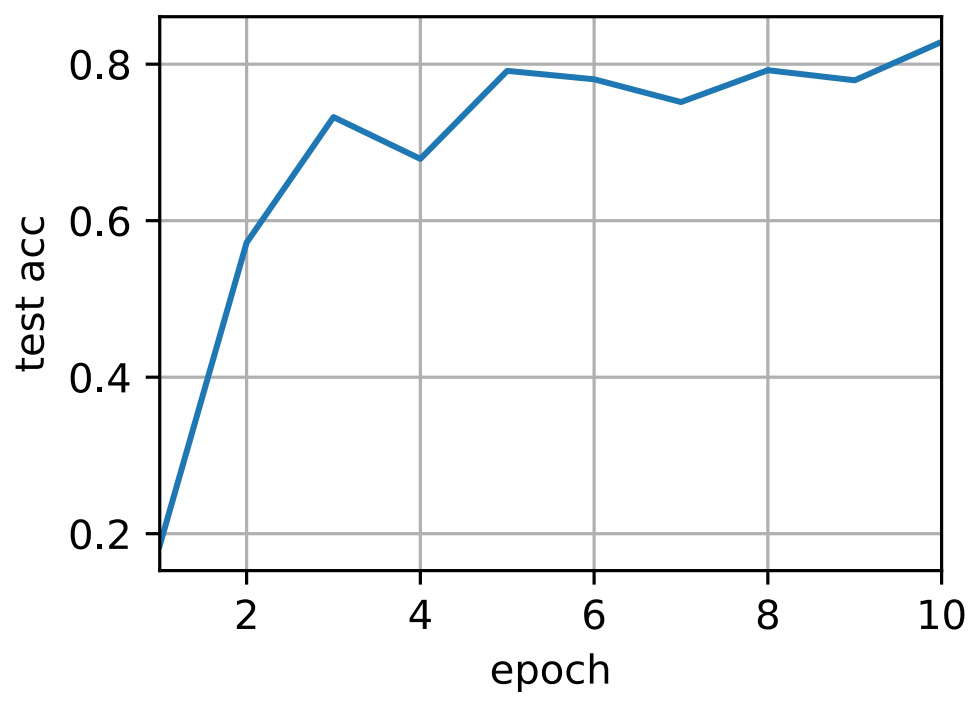

By keeping the batch size and learning rate unchanged and increasing the number of GPUs to 2, we can see that the test accuracy roughly stays the same compared with the previous experiment. In terms of the optimization algorithms, they are identical. Unfortunately there is no meaningful speedup to be gained here: the model is simply too small; moreover we only have a small dataset, where our slightly unsophisticated approach to implementing multi-GPU training suffered from significant Python overhead. We will encounter more complex models and more sophisticated ways of parallelization going forward. Let's see what happens nonetheless for Fashion-MNIST.

```
train(num_gpus=2, batch_size=256, lr=0.2)
```

```
test acc: 0.84, 4.3 sec/epoch on [device(type='cuda', index=0), device(type=
↪'cuda', index=1)]
```

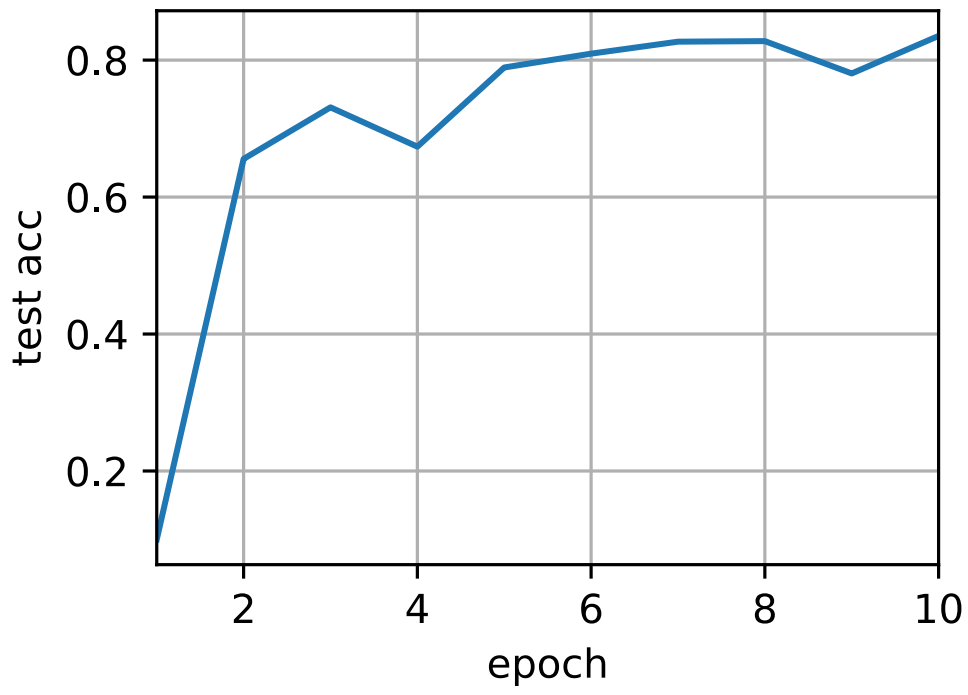

## 13.5.7 Summary

- There are multiple ways to split deep network training over multiple GPUs. We could split them between layers, across layers, or across data. The former two require tightly choreographed data transfers. Data parallelism is the simplest strategy.
- Data parallel training is straightforward. However, it increases the effective minibatch size to be efficient.
- In data parallelism, data is split across multiple GPUs, where each GPU executes its own forward and backward operation and subsequently gradients are aggregated and results are broadcast back to the GPUs.
- We may use slightly increased learning rates for larger minibatches.

## 13.5.8 Exercises

1. When training on $k$ GPUs, change the minibatch size from $b$ to $k \cdot b$, i.e., scale it up by the number of GPUs.
2. Compare accuracy for different learning rates. How does it scale with the number of GPUs?

3. Implement a more efficient `allreduce` function that aggregates different parameters on different GPUs? Why is it more efficient?

4. Implement multi-GPU test accuracy computation.

Discussions[207].

## 13.6 Concise Implementation for Multiple GPUs

Implementing parallelism from scratch for every new model is no fun. Moreover, there is significant benefit in optimizing synchronization tools for high performance. In the following we will show how to do this using high-level APIs of deep learning frameworks. The mathematics and the algorithms are the same as in Section 13.5. Quite unsurprisingly you will need at least two GPUs to run code of this section.

```
import torch
from torch import nn
from d2l import torch as d2l
```

### 13.6.1 A Toy Network

Let's use a slightly more meaningful network than LeNet from Section 13.5 that is still sufficiently easy and quick to train. We pick a ResNet-18 variant (He *et al.*, 2016). Since the input images are tiny we modify it slightly. In particular, the difference from Section 8.6 is that we use a smaller convolution kernel, stride, and padding at the beginning. Moreover, we remove the max-pooling layer.

```
#@save
def resnet18(num_classes, in_channels=1):
    """A slightly modified ResNet-18 model."""
    def resnet_block(in_channels, out_channels, num_residuals,
                     first_block=False):
        blk = []
        for i in range(num_residuals):
            if i == 0 and not first_block:
                blk.append(d2l.Residual(out_channels, use_1x1conv=True,
                                        strides=2))
            else:
                blk.append(d2l.Residual(out_channels))
        return nn.Sequential(*blk)

    # This model uses a smaller convolution kernel, stride, and padding and
    # removes the max-pooling layer
    net = nn.Sequential(
```

```
        nn.Conv2d(in_channels, 64, kernel_size=3, stride=1, padding=1),
        nn.BatchNorm2d(64),
        nn.ReLU())
    net.add_module("resnet_block1", resnet_block(64, 64, 2, first_block=True))
    net.add_module("resnet_block2", resnet_block(64, 128, 2))
    net.add_module("resnet_block3", resnet_block(128, 256, 2))
    net.add_module("resnet_block4", resnet_block(256, 512, 2))
    net.add_module("global_avg_pool", nn.AdaptiveAvgPool2d((1,1)))
    net.add_module("fc", nn.Sequential(nn.Flatten(),
                                       nn.Linear(512, num_classes)))
    return net
```

### 13.6.2 Network Initialization

We will initialize the network inside the training loop. For a refresher on initialization methods see Section 5.4.

```
net = resnet18(10)
# Get a list of GPUs
devices = d2l.try_all_gpus()
# We will initialize the network inside the training loop
```

### 13.6.3 Training

As before, the training code needs to perform several basic functions for efficient parallelism:

- Network parameters need to be initialized across all devices.
- While iterating over the dataset minibatches are to be divided across all devices.
- We compute the loss and its gradient in parallel across devices.
- Gradients are aggregated and parameters are updated accordingly.

In the end we compute the accuracy (again in parallel) to report the final performance of the network. The training routine is quite similar to implementations in previous chapters, except that we need to split and aggregate data.

```
def train(net, num_gpus, batch_size, lr):
    train_iter, test_iter = d2l.load_data_fashion_mnist(batch_size)
    devices = [d2l.try_gpu(i) for i in range(num_gpus)]
    def init_weights(module):
        if type(module) in [nn.Linear, nn.Conv2d]:
            nn.init.normal_(module.weight, std=0.01)
    net.apply(init_weights)
    # Set the model on multiple GPUs
    net = nn.DataParallel(net, device_ids=devices)
```

```
    trainer = torch.optim.SGD(net.parameters(), lr)
    loss = nn.CrossEntropyLoss()
    timer, num_epochs = d2l.Timer(), 10
    animator = d2l.Animator('epoch', 'test acc', xlim=[1, num_epochs])
    for epoch in range(num_epochs):
        net.train()
        timer.start()
        for X, y in train_iter:
            trainer.zero_grad()
            X, y = X.to(devices[0]), y.to(devices[0])
            l = loss(net(X), y)
            l.backward()
            trainer.step()
        timer.stop()
        animator.add(epoch + 1, (d2l.evaluate_accuracy_gpu(net, test_iter),))
    print(f'test acc: {animator.Y[0][-1]:.2f}, {timer.avg():.1f} sec/epoch '
          f'on {str(devices)}')
```

Let's see how this works in practice. As a warm-up we train the network on a single GPU.

```
train(net, num_gpus=1, batch_size=256, lr=0.1)
```

```
test acc: 0.92, 13.1 sec/epoch on [device(type='cuda', index=0)]
```

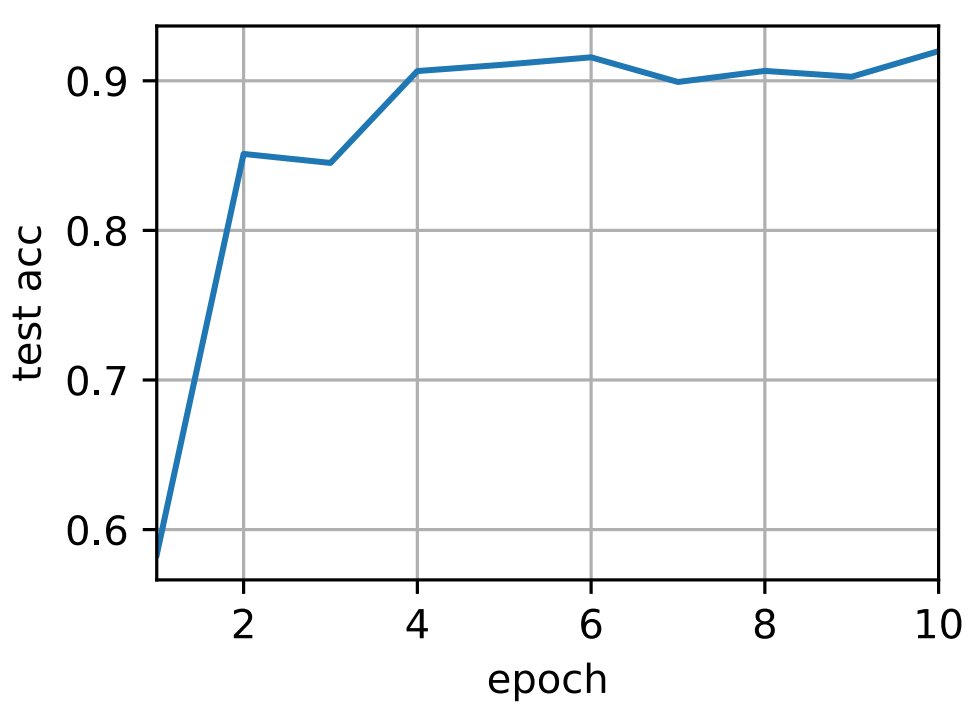

Next we use 2 GPUs for training. Compared with LeNet evaluated in Section 13.5, the model for ResNet-18 is considerably more complex. This is where parallelization shows its advantage. The time for computation is meaningfully larger than the time for synchronizing parameters. This improves scalability since the overhead for parallelization is less relevant.

```
train(net, num_gpus=2, batch_size=512, lr=0.2)
```

```
test acc: 0.79, 10.6 sec/epoch on [device(type='cuda', index=0), device(type=
↪'cuda', index=1)]
```

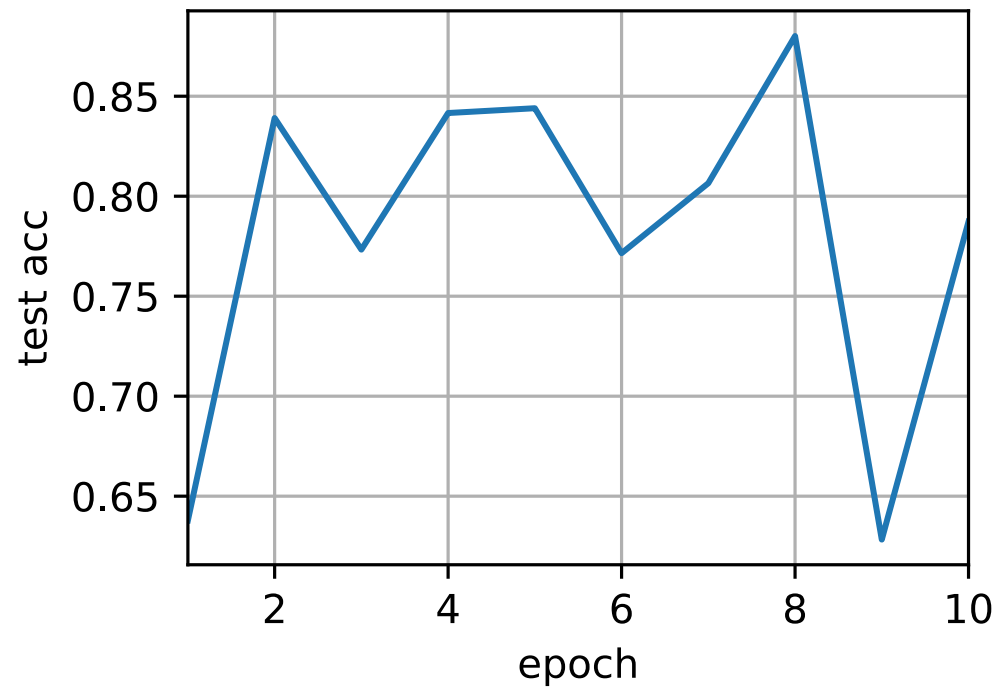

### 13.6.4 Summary

- Data is automatically evaluated on the devices where the data can be found.
- Take care to initialize the networks on each device before trying to access the parameters on that device. Otherwise you will encounter an error.
- The optimization algorithms automatically aggregate over multiple GPUs.

### 13.6.5 Exercises

1. This section uses ResNet-18. Try different epochs, batch sizes, and learning rates. Use more GPUs for computation. What happens if you try this with 16 GPUs (e.g., on an AWS p2.16xlarge instance)?
2. Sometimes, different devices provide different computing power. We could use the GPUs and the CPU at the same time. How should we divide the work? Is it worth the effort? Why? Why not?

Discussions[208].

## 13.7 Parameter Servers

As we move from a single GPU to multiple GPUs and then to multiple servers containing multiple GPUs, possibly all spread out across multiple racks and network switches, our algorithms for distributed and parallel training need to become much more sophisticated. Details matter since different interconnects have very different bandwidth (e.g., NVLink can offer up to 100 GB/s across 6 links in an appropriate setting, PCIe 4.0 (16-lane) offers 32 GB/s, while even high speed 100GbE Ethernet only amounts to 10 GB/s). At the same time it is unreasonable to expect that a statistical modeler be an expert in networking and systems.

The core idea of the parameter server was introduced in Smola and Narayanamurthy (2010) in the context of distributed latent variable models. A description of the push and pull semantics then followed in Ahmed *et al.* (2012) and a description of the system and an open source library followed in Li *et al.* (2014). In the following we will motivate the components needed for efficiency.

## 13.7.1 Data-Parallel Training

Let's review the data parallel training approach to distributed training. We will use this to the exclusion of all others in this section since it is significantly simpler to implement in practice. There are virtually no use cases (besides deep learning on graphs) where any other strategy for parallelism is preferred since GPUs have plenty of memory nowadays. Fig. 13.7.1 describes the variant of data parallelism that we implemented in Section 13.5. The key aspect in it is that the aggregation of gradients occurs on one single GPU (GPU 0) before the updated parameters are rebroadcast to all GPUs.

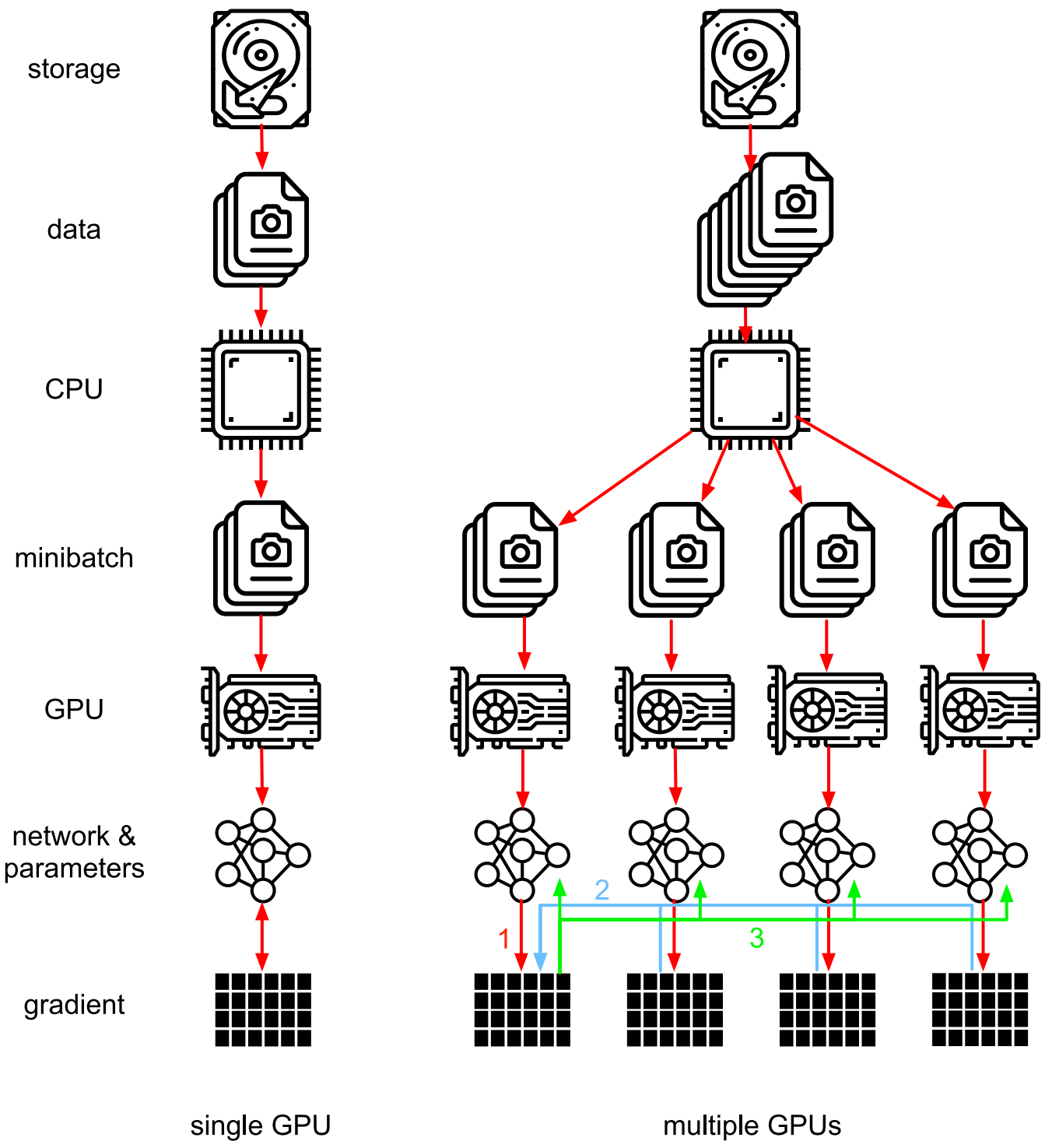

Fig. 13.7.1 Left: single GPU training. Right: a variant of multi-GPU training: (1) we compute loss and gradient, (2) all gradients are aggregated on one GPU, (3) parameter update happens and the parameters are re-distributed to all GPUs.

In retrospect, the decision to aggregate on GPU 0 seems rather ad-hoc. After all, we might just as well aggregate on the CPU. In fact, we could even decide to aggregate some of the parameters on one GPU and some others on another. Provided that the optimization algorithm supports this, there is no real reason for why we could not. For instance, if we have four parameter vectors with associated gradients $\mathbf{g}_1, \ldots, \mathbf{g}_4$ we could aggregate the gradients on one GPU for each $\mathbf{g}_i$ $(i = 1, \ldots, 4)$.

This reasoning seems arbitrary and frivolous. After all, the mathematics is the same throughout. However, we are dealing with real physical hardware where different buses have different bandwidth as discussed in Section 13.4. Consider a real 4-way GPU server as described in Fig. 13.7.2. If it is particularly well connected, it might have a 100 GbE network card. More typical numbers are in the 1–10 GbE range with an effective bandwidth of 100 MB/s to 1 GB/s. Since the CPUs have too few PCIe lanes to connect to all GPUs directly (e.g., consumer-grade Intel CPUs have 24 lanes) we need a multiplexer[209]. The bandwidth from the CPU on a 16x Gen3 link is 16 GB/s. This is also the speed at which *each* of the GPUs is connected to the switch. This means that it is more effective to communicate between the devices.

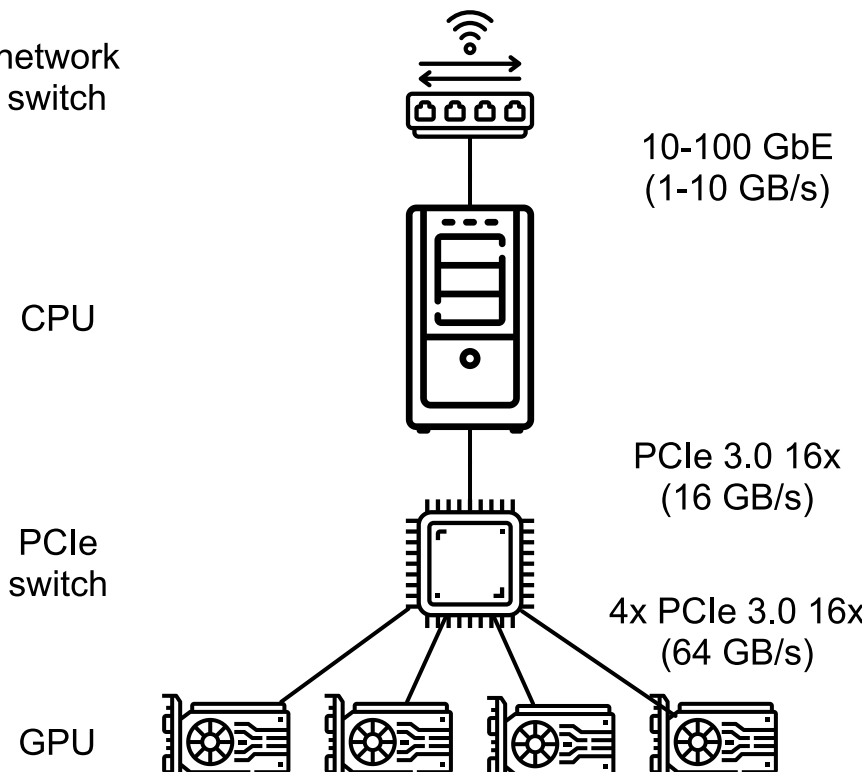

Fig. 13.7.2 A 4-way GPU server.

For the sake of the argument let's assume that the gradients are of 160 MB. In this case it takes 30 ms to send the gradients from all 3 remaining GPUs to the fourth one (each transfer takes 10 ms = 160 MB / 16 GB/s). Adding another 30 ms to transmit the weight vectors back we arrive at a total of 60 ms. If we send all data to the CPU we incur a penalty of 40 ms since *each* of the four GPUs needs to send the data to the CPU, yielding a total of 80 ms. Lastly assume that we are able to split the gradients into 4 parts of 40 MB each. Now we can aggregate each of the parts on a different GPU *simultaneously* since the PCIe switch offers a full-bandwidth operation between all links. Instead of 30 ms this takes 7.5 ms, yielding a total of 15 ms for a synchronization operation. In short, depending on how we synchronize parameters the same operation can take anywhere from 15 ms to 80 ms. Fig. 13.7.3 depicts the different strategies for exchanging parameters.

Note that we have yet another tool at our disposal when it comes to improving performance: in a deep network it takes some time to compute all gradients from the top to the bottom. We can begin synchronizing gradients for some parameter groups even while we are still busy computing them for others. See e.g., Sergeev and Del Balso (2018) for details on how to do this in Horovod[210].

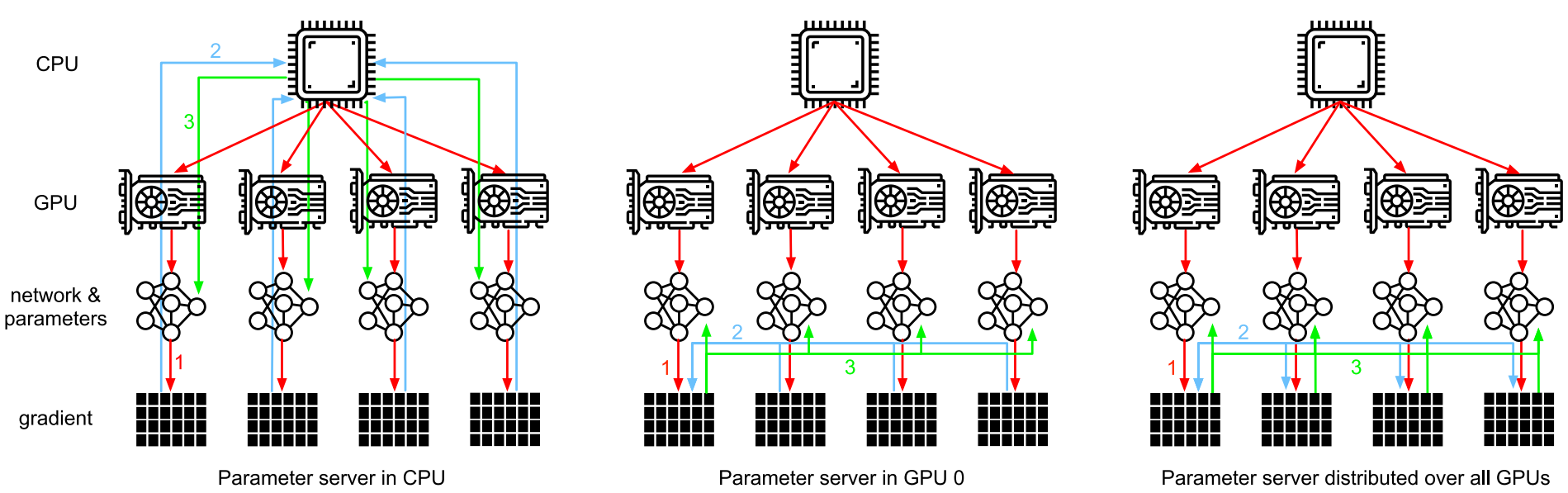

Fig. 13.7.3 Parameter synchronization strategies.

## 13.7.2 Ring Synchronization

When it comes to synchronization on modern deep learning hardware we often encounter significantly bespoke network connectivity. For instance, the AWS p3.16xlarge and NVIDIA DGX-2 instances share the connectivity structure of Fig. 13.7.4. Each GPU connects to a host CPU via a PCIe link which operates at best at 16 GB/s. Additionally each GPU also has 6 NVLink connections, each of which is capable of transferring 300 Gbit/s bidirectionally. This amounts to around 18 GB/s per link per direction. In short, the aggregate NVLink bandwidth is significantly higher than the PCIe bandwidth. The question is how to use it most efficiently.

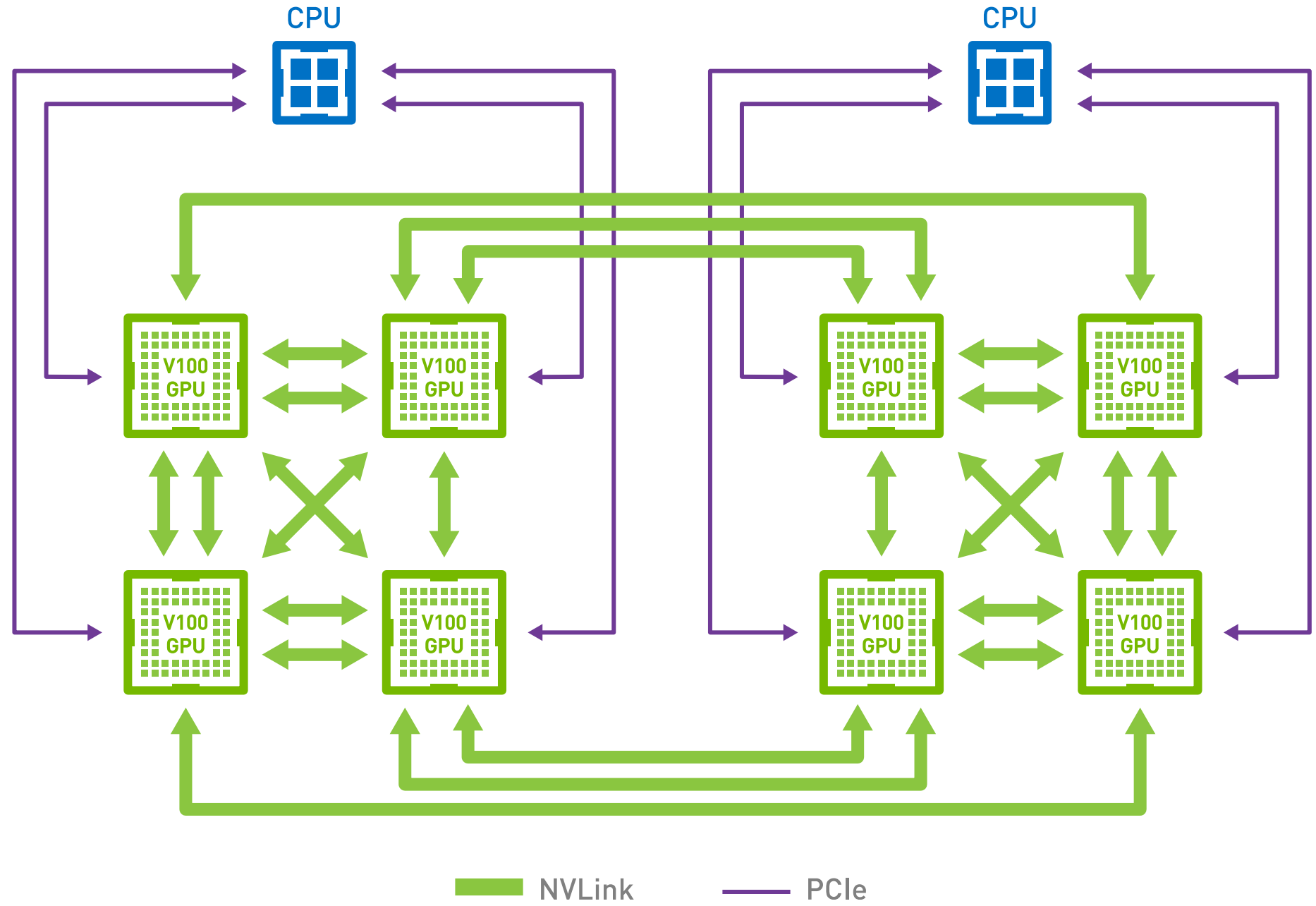

Fig. 13.7.4 NVLink connectivity on 8 V100 GPU servers (image courtesy of NVIDIA).

It turns out that the optimal synchronization strategy is to decompose the network into two rings and to use them to synchronize data directly (Wang *et al.*, 2018). Fig. 13.7.5 illustrates that the network can be decomposed into one ring (1-2-3-4-5-6-7-8-1) with double NVLink bandwidth and into one (1-4-6-3-5-8-2-7-1) with regular bandwidth. Designing an efficient synchronization protocol in this case is nontrivial.

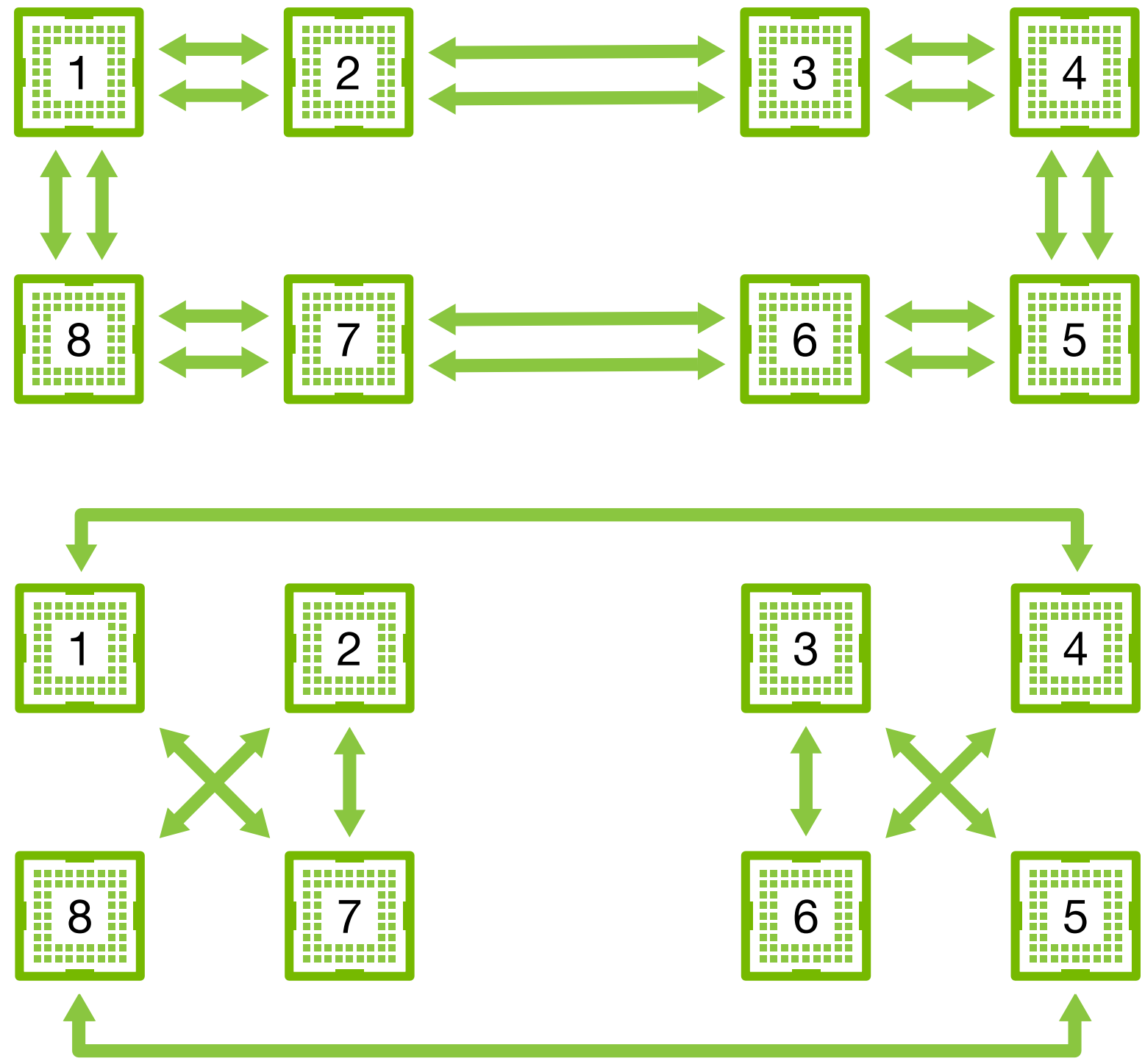

Fig. 13.7.5 Decomposition of the NVLink network into two rings.

Consider the following thought experiment: given a ring of $n$ computing nodes (or GPUs) we can send gradients from the first to the second node. There it is added to the local gradient and sent on to the third node, and so on. After $n-1$ steps the aggregate gradient can be found in the last-visited node. That is, the time to aggregate gradients grows linearly with the number of nodes. But if we do this the algorithm is quite inefficient. After all, at any time there is only one of the nodes communicating. What if we broke the gradients into $n$ chunks and started synchronizing chunk $i$ starting at node $i$? Since each chunk is of size $1/n$ the total time is now $(n-1)/n \approx 1$. In other words, the time spent to aggregate gradients *does not grow* as we increase the size of the ring. This is quite an astonishing result. Fig. 13.7.6 illustrates the sequence of steps on $n = 4$ nodes.

If we use the same example of synchronizing 160 MB across 8 V100 GPUs we arrive at approximately $2 \cdot 160\text{MB}/(3 \cdot 18\text{GB/s}) \approx 6\text{ms}$. This is better than using the PCIe bus, even though we are now using 8 GPUs. Note that in practice these numbers are a bit worse, since deep learning frameworks often fail to assemble communication into large burst transfers.

Note that there is a common misconception that ring synchronization is fundamentally dif-ferent from other synchronization algorithms. The only difference is that the synchronization path is somewhat more elaborate when compared with a simple tree.

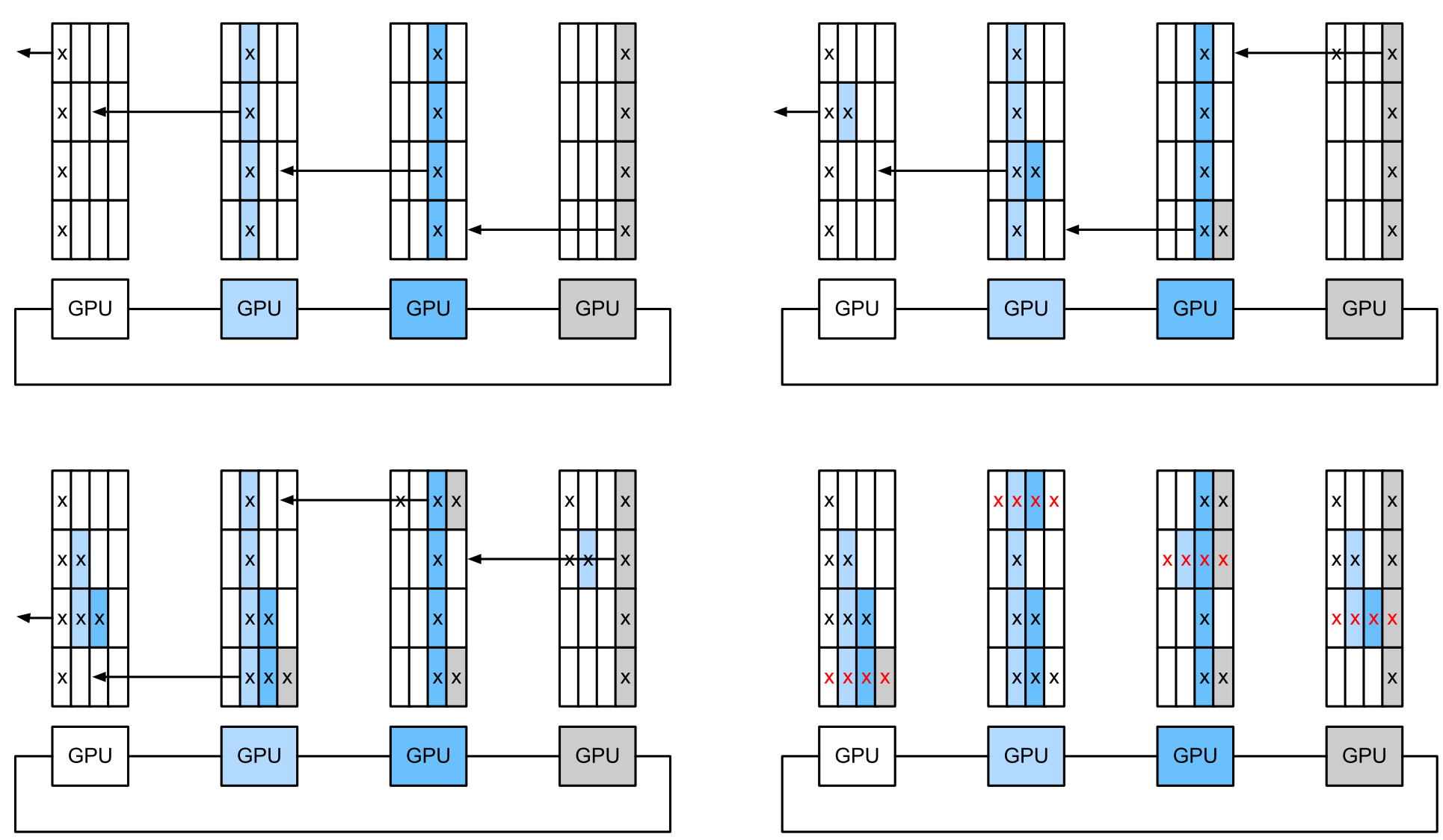

Fig. 13.7.6 Ring synchronization across 4 nodes. Each node starts transmitting parts of gradients to its left neighbor until the assembled gradient can be found in its right neighbor.

## 13.7.3 Multi-Machine Training

Distributed training on multiple machines adds a further challenge: we need to communicate with servers that are only connected across a comparatively lower bandwidth fabric that can be over an order of magnitude slower in some cases. Synchronization across devices is tricky. After all, different machines running training code will have subtly different speed. Hence we need to *synchronize* them if we want to use synchronous distributed optimization. Fig. 13.7.7 illustrates how distributed parallel training occurs.

1. A (different) batch of data is read on each machine, split across multiple GPUs and transferred to GPU memory. There predictions and gradients are computed on each GPU batch separately.
2. The gradients from all local GPUs are aggregated on one GPU (or parts of it are aggregated over different GPUs).
3. The gradients are sent to the CPUs.
4. The CPUs send the gradients to a central parameter server which aggregates all the gradients.
5. The aggregate gradients are then used to update the parameters and the updated parameters are broadcast back to the individual CPUs.

6. The information is sent to one (or multiple) GPUs.
7. The updated parameters are spread across all GPUs.

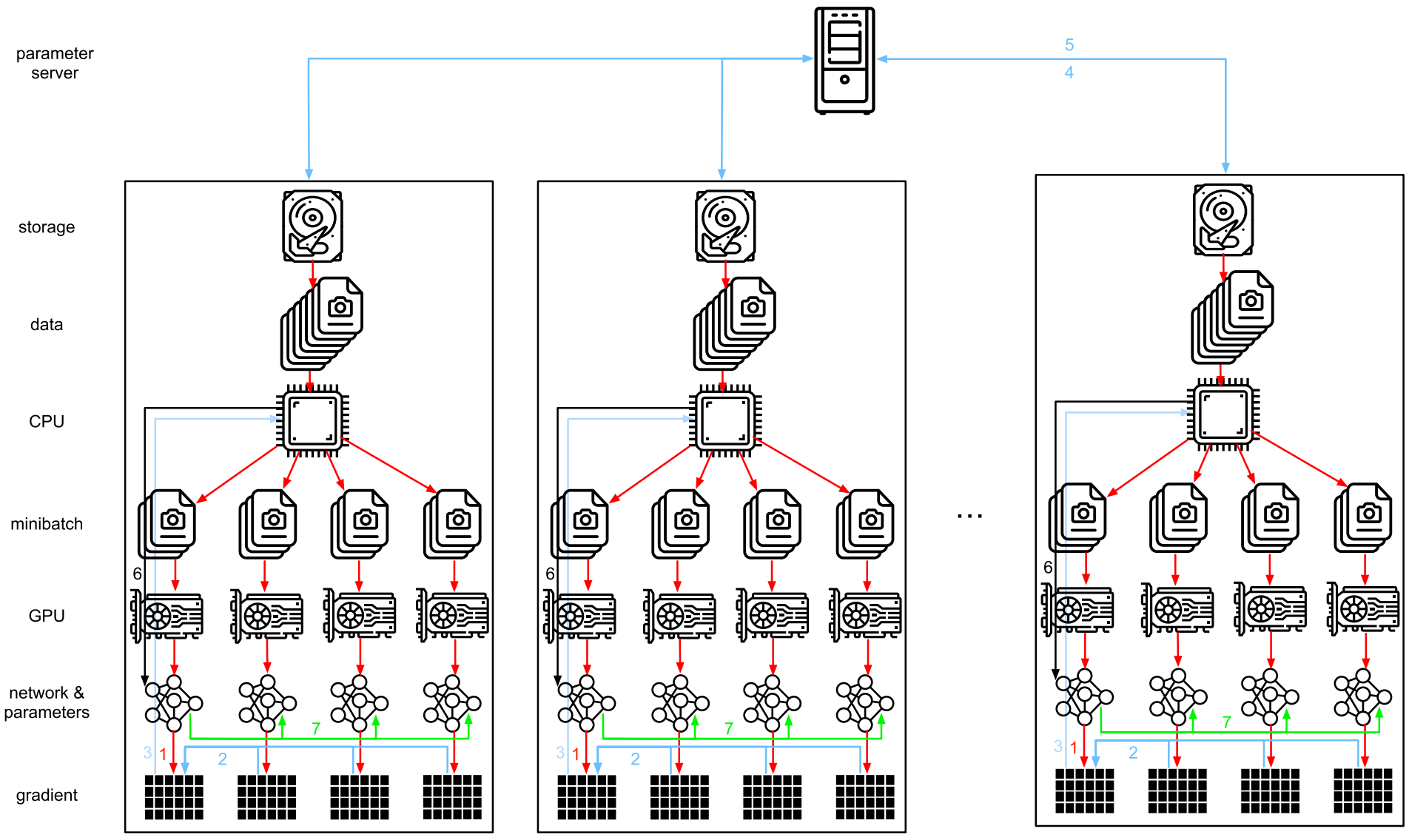

Fig. 13.7.7 Multi-machine multi-GPU distributed parallel training.

Each of these operations seems rather straightforward. And, indeed, they can be carried out efficiently *within* a single machine. Once we look at multiple machines, though, we can see that the central parameter server becomes the bottleneck. After all, the bandwidth per server is limited, hence for $m$ workers the time it takes to send all gradients to the server is $O(m)$. We can break through this barrier by increasing the number of servers to $n$. At this point each server only needs to store $O(1/n)$ of the parameters, hence the total time for updates and optimization becomes $O(m/n)$. Matching both numbers yields constant scaling regardless of how many workers we are dealing with. In practice we use the *same* machines both as workers and as servers. Fig. 13.7.8 illustrates the design (see also (Li *et al.*, 2014) for details). In particular, ensuring that multiple machines work without unreasonable delays is nontrivial.

## 13.7.4 Key–Value Stores

Implementing the steps required for distributed multi-GPU training in practice is nontrivial. This is why it pays to use a common abstraction, namely that of a *key–value store* with redefined update semantics.

Across many workers and many GPUs the computation for gradient $i$ can be defined as

$$\mathbf{g}_i = \sum_{k \in \text{workers}} \sum_{j \in \text{GPUs}} \mathbf{g}_{ijk}, \tag{13.7.1}$$

where $\mathbf{g}_{ijk}$ is part of gradient $i$ split on GPU $j$ of worker $k$. The key aspect in this operation is that it is a *commutative reduction*, that is, it turns many vectors into one and the order in which the operation is applied does not matter. This is great for our purposes since we do not (need to) have fine grained control over when which gradient is received. Besides, note that this operation is independent among different $i$.

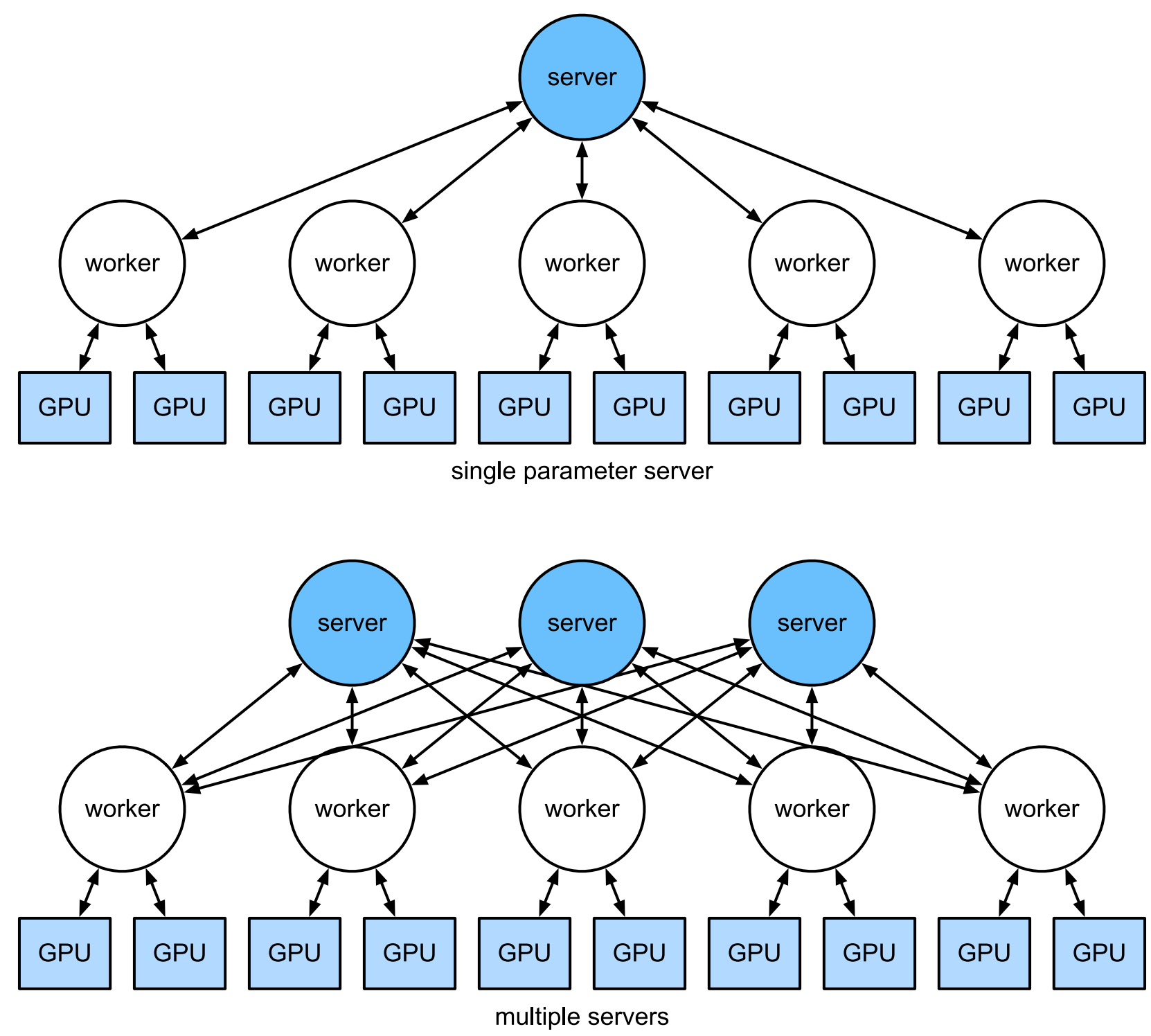

Fig. 13.7.8 Top: a single parameter server is a bottleneck since its bandwidth is finite. Bottom: multiple parameter servers store parts of the parameters with aggregate bandwidth.

This allows us to define the following two operations: *push*, which accumulates gradients, and *pull*, which retrieves aggregate gradients. Since we have many different sets of gradients (after all, we have many layers), we need to index the gradients with a key $i$. This similarity to key–value stores, such as the one introduced in Dynamo (DeCandia *et al.*, 2007) is not by coincidence. They, too, satisfy many similar characteristics, in particular when it comes to distributing the parameters across multiple servers.

The push and pull operations for key-value stores are described as follows:

- **push(key, value)** sends a particular gradient (the value) from a worker to a common storage. There the value is aggregated, e.g., by summing it up.
- **pull(key, value)** retrieves an aggregate value from common storage, e.g., after combining the gradients from all workers.

By hiding all the complexity about synchronization behind a simple push and pull operation we can decouple the concerns of statistical modelers who want to be able to express opti-mization in simple terms and the system engineers who need to deal with the complexity inherent in distributed synchronization.

### 13.7.5 Summary

- Synchronization needs to be highly adaptive to specific network infrastructure and connectivity within a server. This can make a significant difference to the time it takes to synchronize.
- Ring-synchronization can be optimal for p3 and DGX-2 servers. For others possibly not so much.
- A hierarchical synchronization strategy works well when adding multiple parameter servers for increased bandwidth.

### 13.7.6 Exercises

1. Can you increase the ring synchronization even further? Hint: you can send messages in both directions.
2. Is it possible to allow asynchronous communication (while computation is still ongoing)? How does it affect performance?
3. What if we lost a server during a long-running computation? How can we design a *fault tolerance* mechanism to avoid restarting the computation fully?

Discussions[211].

# 14 Computer Vision

Whether it is medical diagnosis, self-driving vehicles, camera monitoring, or smart filters, many applications in the field of computer vision are closely related to our current and future lives. In recent years, deep learning has been the transformative power for advancing the performance of computer vision systems. It can be said that the most advanced computer vision applications are almost inseparable from deep learning. In view of this, this chapter will focus on the field of computer vision, and investigate methods and applications that have recently been influential in academia and industry.

In Chapter 7 and Chapter 8, we studied various convolutional neural networks that are commonly used in computer vision, and applied them to simple image classification tasks. At the beginning of this chapter, we will describe two methods that may improve model generalization, namely *image augmentation* and *fine-tuning*, and apply them to image classification. Since deep neural networks can effectively represent images in multiple levels, such layerwise representations have been successfully used in various computer vision tasks such as *object detection*, *semantic segmentation*, and *style transfer*. Following the key idea of leveraging layerwise representations in computer vision, we will begin with major components and techniques for object detection. Next, we will show how to use *fully convolutional networks* for semantic segmentation of images. Then we will explain how to use style transfer techniques to generate images like the cover of this book. In the end, we conclude this chapter by applying the materials of this chapter and several previous chapters on two popular computer vision benchmark datasets.

## 14.1 Image Augmentation

In Section 8.1, we mentioned that large datasets are a prerequisite for the success of deep neural networks in various applications. *Image augmentation* generates similar but distinct training examples after a series of random changes to the training images, thereby expanding the size of the training set. Alternatively, image augmentation can be motivated by the fact that random tweaks of training examples allow models to rely less on certain attributes, thereby improving their generalization ability. For example, we can crop an image in different ways to make the object of interest appear in different positions, thereby reducing the dependence of a model on the position of the object. We can also adjust factors such as brightness and color to reduce a model's sensitivity to color. It is probably true that image augmentation was indispensable for the success of AlexNet at that time. In this section we will discuss this widely used technique in computer vision.

```
%matplotlib inline
import torch
import torchvision
from torch import nn
from d2l import torch as d2l
```

## 14.1.1 Common Image Augmentation Methods

In our investigation of common image augmentation methods, we will use the following $400 \times 500$ image an example.

```
d2l.set_figsize()
img = d2l.Image.open('../img/cat1.jpg')
d2l.plt.imshow(img);
```

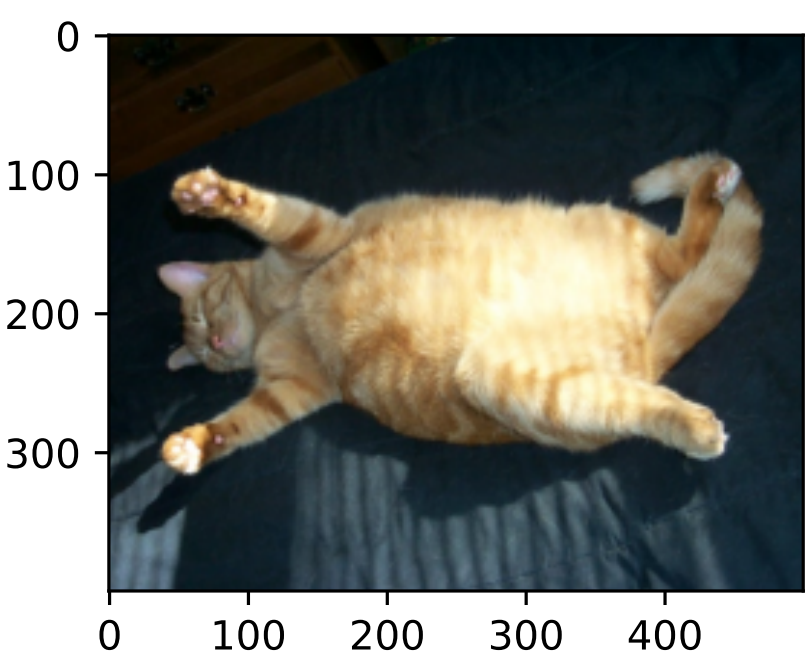

Most image augmentation methods have a certain degree of randomness. To make it easier for us to observe the effect of image augmentation, next we define an auxiliary function `apply`. This function runs the image augmentation method `aug` multiple times on the input image `img` and shows all the results.

```
def apply(img, aug, num_rows=2, num_cols=4, scale=1.5):
    Y = [aug(img) for _ in range(num_rows * num_cols)]
    d2l.show_images(Y, num_rows, num_cols, scale=scale)
```

### Flipping and Cropping

Flipping the image left and right usually does not change the category of the object. This is one of the earliest and most widely used methods of image augmentation. Next, we use the

`transforms` module to create the `RandomHorizontalFlip` instance, which flips an image left and right with a 50% chance.

```
apply(img, torchvision.transforms.RandomHorizontalFlip())
```

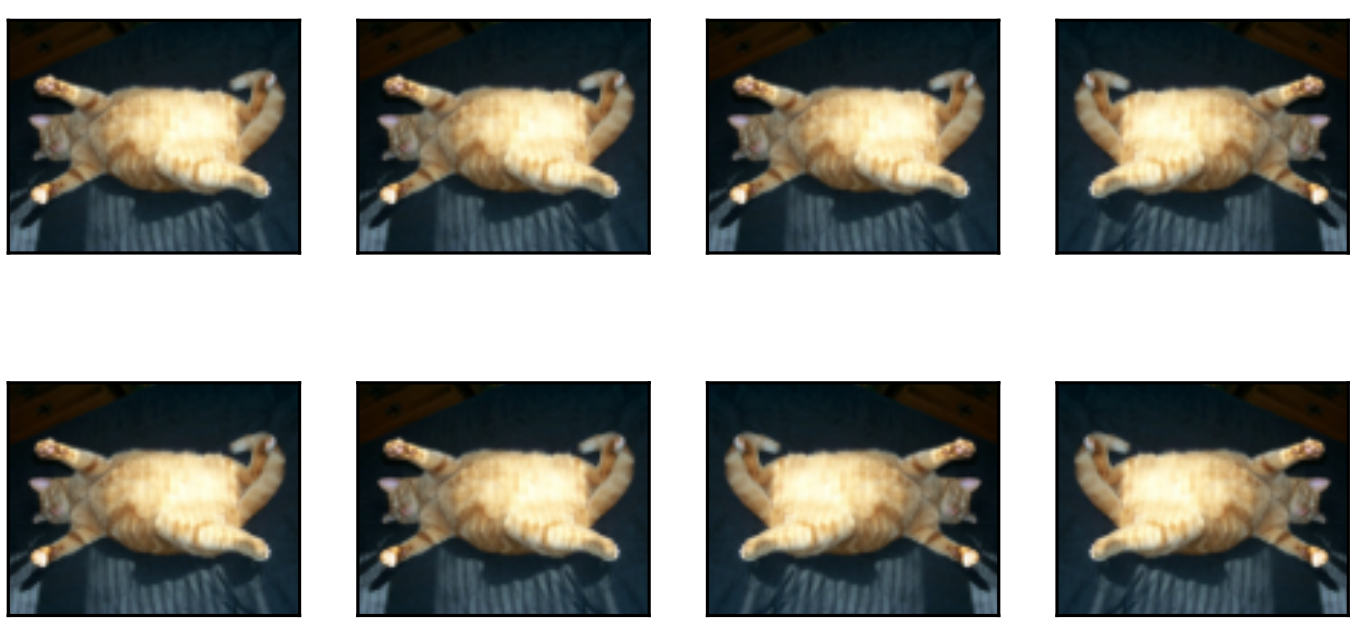

Flipping up and down is not as common as flipping left and right. But at least for this example image, flipping up and down does not hinder recognition. Next, we create a `RandomVerticalFlip` instance to flip an image up and down with a 50% chance.

```
apply(img, torchvision.transforms.RandomVerticalFlip())
```

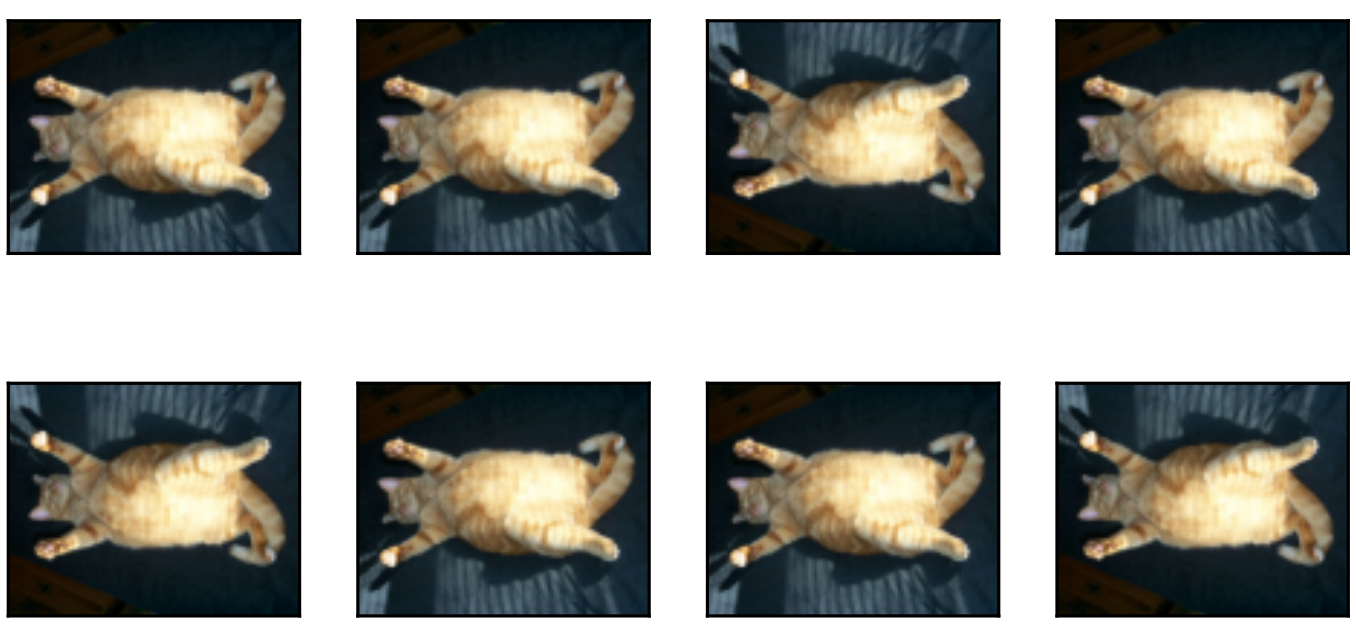

In the example image we used, the cat is in the middle of the image, but this may not be the case in general. In Section 7.5, we explained that the pooling layer can reduce the sensitivity of a convolutional layer to the target position. In addition, we can also randomly crop the image to make objects appear in different positions in the image at different scales, which can also reduce the sensitivity of a model to the target position.

In the code below, we randomly crop an area with an area of $10\% \sim 100\%$ of the original area each time, and the ratio of width to height of this area is randomly selected from $0.5 \sim 2$. Then, the width and height of the region are both scaled to 200 pixels. Unless otherwise specified, the random number between $a$ and $b$ in this section refers to a continuous value obtained by random and uniform sampling from the interval $[a, b]$.

```
shape_aug = torchvision.transforms.RandomResizedCrop(
    (200, 200), scale=(0.1, 1), ratio=(0.5, 2))
apply(img, shape_aug)
```

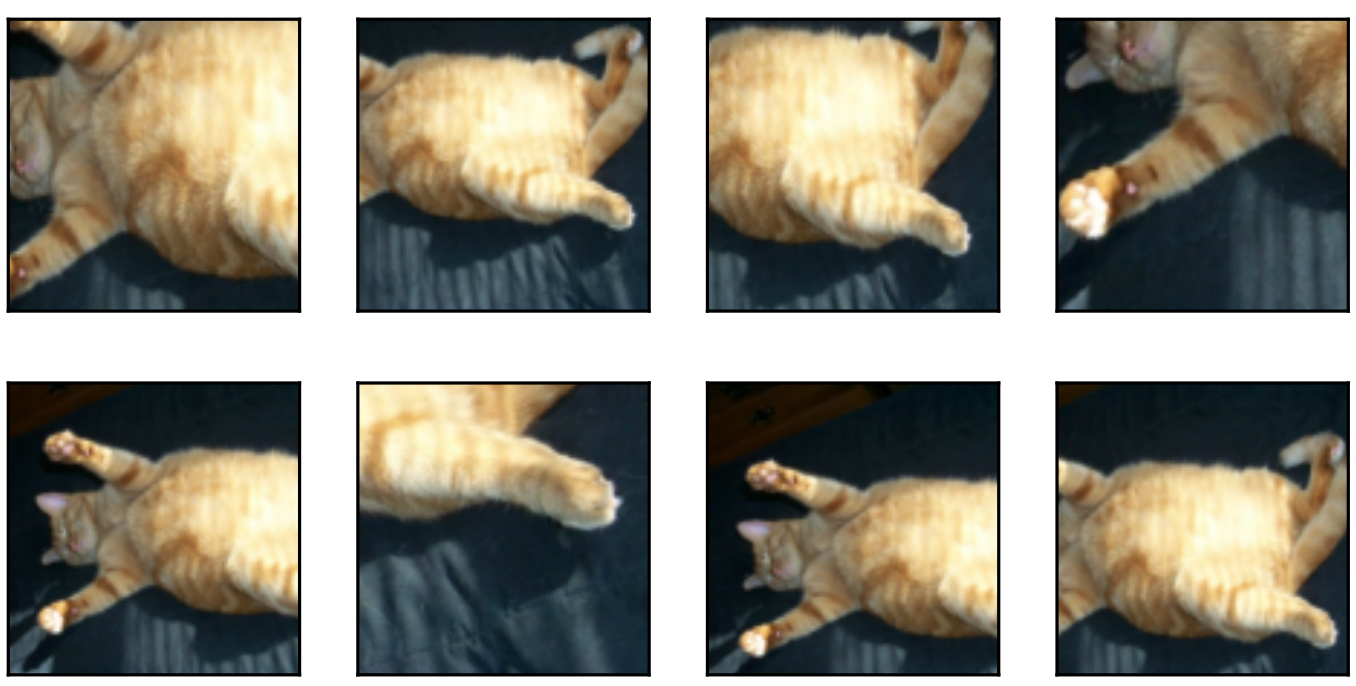

### Changing Colors

Another augmentation method is changing colors. We can change four aspects of the image color: brightness, contrast, saturation, and hue. In the example below, we randomly change the brightness of the image to a value between 50% ($1 - 0.5$) and 150% ($1 + 0.5$) of the original image.

```
apply(img, torchvision.transforms.ColorJitter(
    brightness=0.5, contrast=0, saturation=0, hue=0))
```

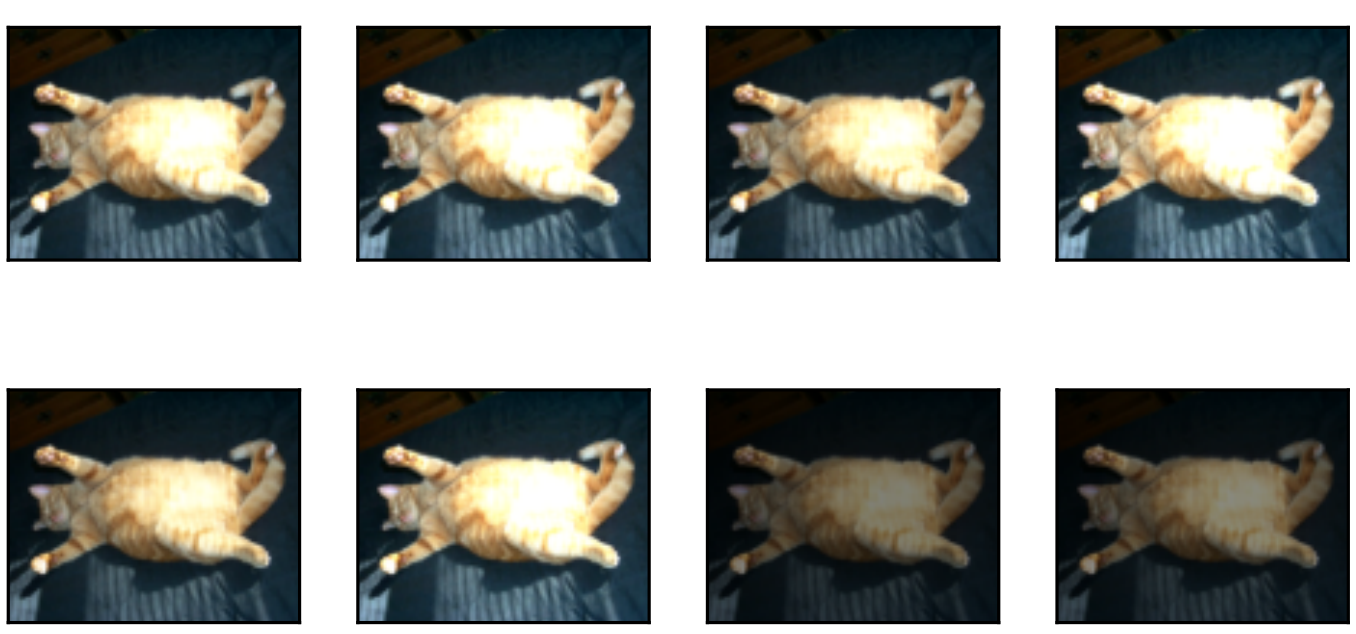

Similarly, we can randomly change the hue of the image.

```
apply(img, torchvision.transforms.ColorJitter(
    brightness=0, contrast=0, saturation=0, hue=0.5))
```

We can also create a `RandomColorJitter` instance and set how to randomly change the `brightness`, `contrast`, `saturation`, and hue of the image at the same time.

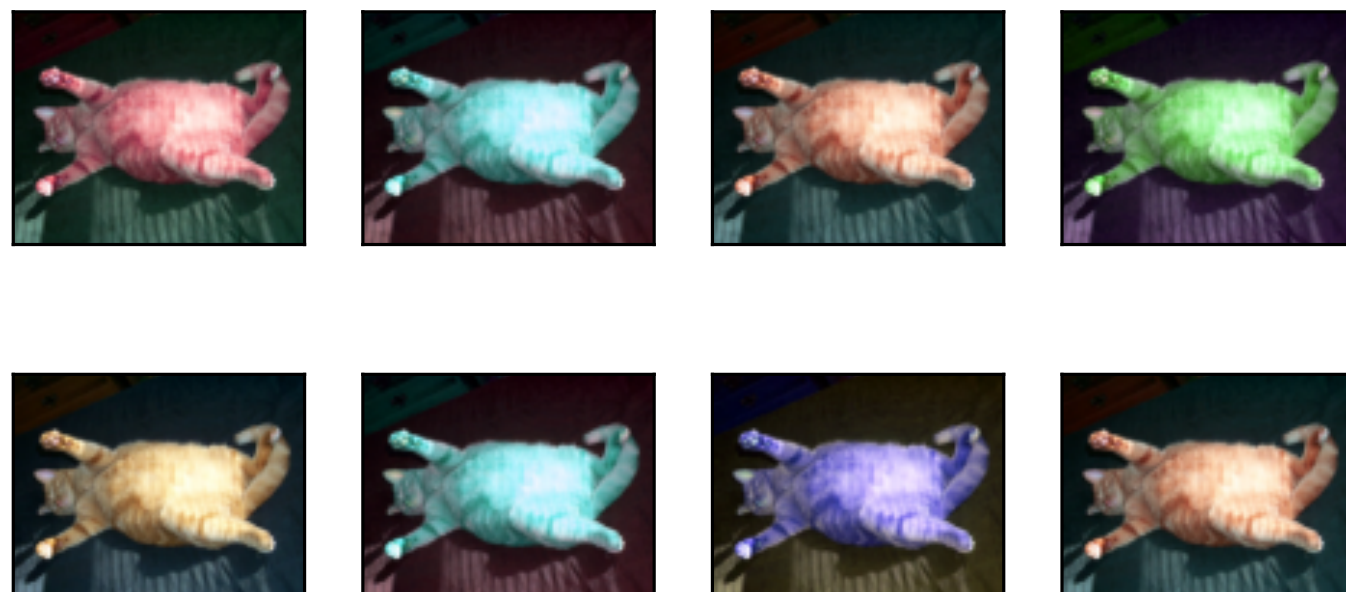

```
color_aug = torchvision.transforms.ColorJitter(
    brightness=0.5, contrast=0.5, saturation=0.5, hue=0.5)
apply(img, color_aug)
```

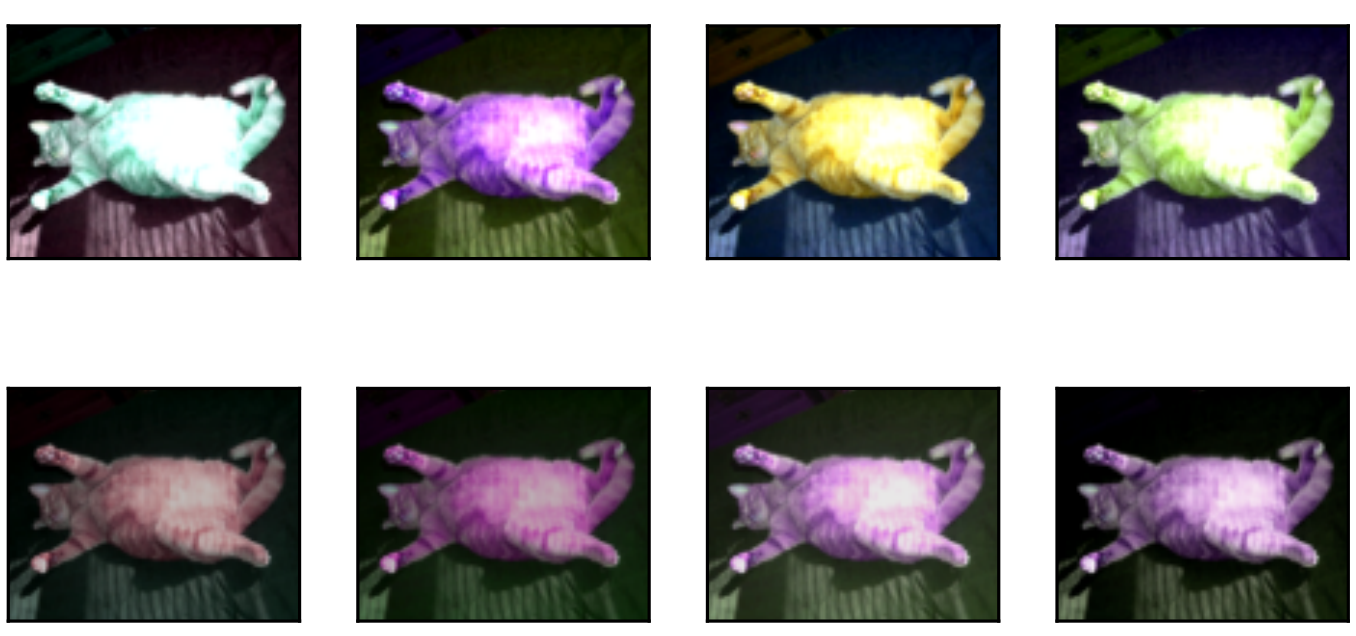

### Combining Multiple Image Augmentation Methods

In practice, we will combine multiple image augmentation methods. For example, we can combine the different image augmentation methods defined above and apply them to each image via a `Compose` instance.

```
augs = torchvision.transforms.Compose([
    torchvision.transforms.RandomHorizontalFlip(), color_aug, shape_aug])
apply(img, augs)
```

## 14.1.2 Training with Image Augmentation

Let's train a model with image augmentation. Here we use the CIFAR-10 dataset instead of the Fashion-MNIST dataset that we used before. This is because the position and size of the objects in the Fashion-MNIST dataset have been normalized, while the color and size of the objects in the CIFAR-10 dataset have more significant differences. The first 32 training images in the CIFAR-10 dataset are shown below.

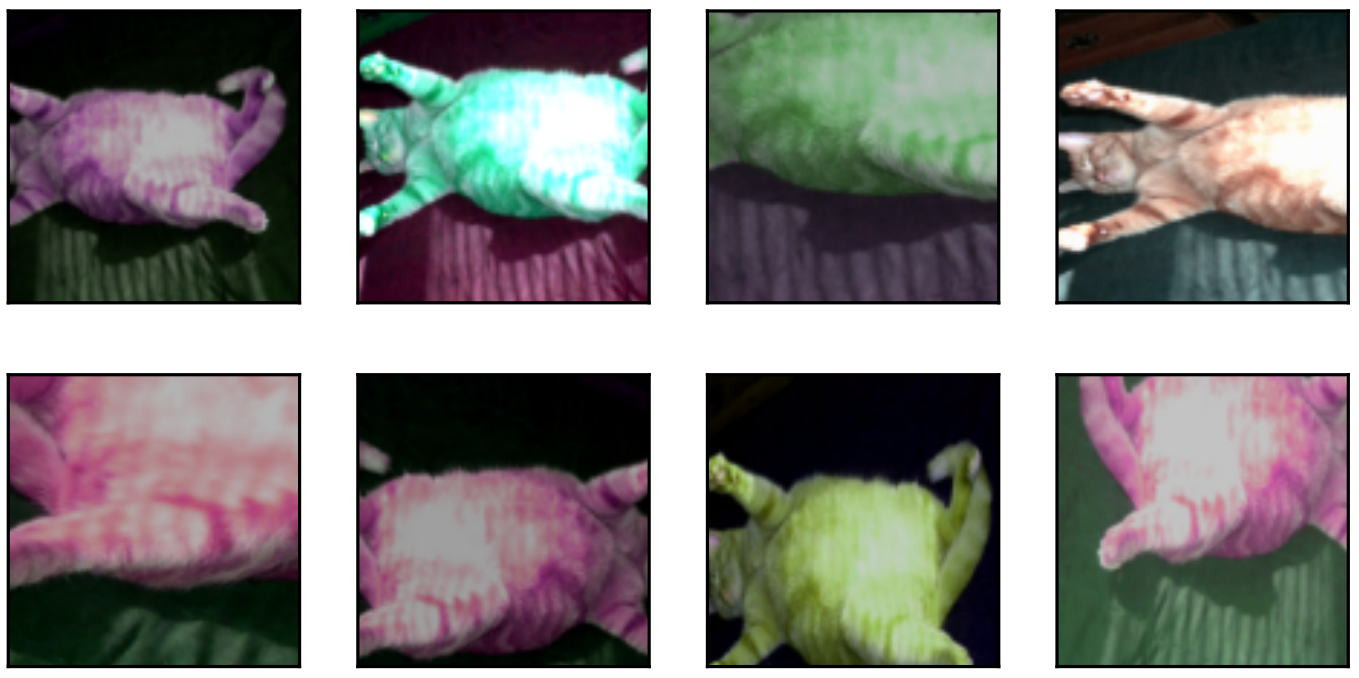

```
all_images = torchvision.datasets.CIFAR10(train=True, root="../data",
                                          download=True)
d2l.show_images([all_images[i][0] for i in range(32)], 4, 8, scale=0.8);
```

```
Files already downloaded and verified
```

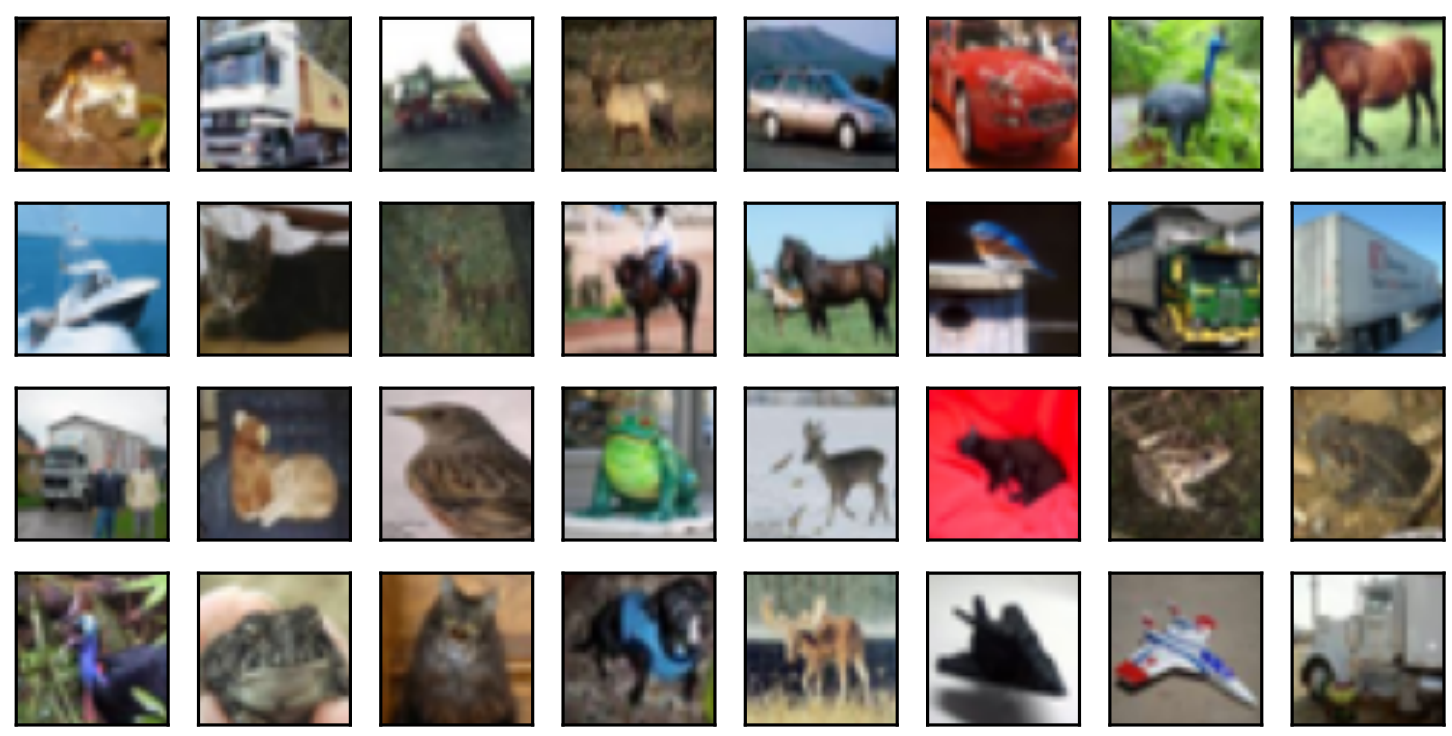

In order to obtain definitive results during prediction, we usually only apply image augmentation to training examples, and do not use image augmentation with random operations during prediction. Here we only use the simplest random left-right flipping method. In addition, we use a `ToTensor` instance to convert a minibatch of images into the format required by the deep learning framework, i.e., 32-bit floating point numbers between 0 and 1 with the shape of (batch size, number of channels, height, width).

```
train_augs = torchvision.transforms.Compose([
     torchvision.transforms.RandomHorizontalFlip(),
     torchvision.transforms.ToTensor()])

test_augs = torchvision.transforms.Compose([
     torchvision.transforms.ToTensor()])
```

Next, we define an auxiliary function to facilitate reading the image and applying image aug-mentation. The `transform` argument provided by PyTorch's dataset applies augmentation to transform the images. For a detailed introduction to `DataLoader`, please refer to Section 4.2.

```
def load_cifar10(is_train, augs, batch_size):
    dataset = torchvision.datasets.CIFAR10(root="../data", train=is_train,
                                           transform=augs, download=True)
    dataloader = torch.utils.data.DataLoader(dataset, batch_size=batch_size,
                    shuffle=is_train, num_workers=d2l.get_dataloader_workers())
    return dataloader
```

## Multi-GPU Training

We train the ResNet-18 model from Section 8.6 on the CIFAR-10 dataset. Recall the introduction to multi-GPU training in Section 13.6. In the following, we define a function to train and evaluate the model using multiple GPUs.

```
#@save
def train_batch_ch13(net, X, y, loss, trainer, devices):
    """Train for a minibatch with multiple GPUs (defined in Chapter 13)."""
    if isinstance(X, list):
        # Required for BERT fine-tuning (to be covered later)
        X = [x.to(devices[0]) for x in X]
    else:
        X = X.to(devices[0])
    y = y.to(devices[0])
    net.train()
    trainer.zero_grad()
    pred = net(X)
    l = loss(pred, y)
    l.sum().backward()
    trainer.step()
    train_loss_sum = l.sum()
    train_acc_sum = d2l.accuracy(pred, y)
    return train_loss_sum, train_acc_sum
```

```
#@save
def train_ch13(net, train_iter, test_iter, loss, trainer, num_epochs,
               devices=d2l.try_all_gpus()):
    """Train a model with multiple GPUs (defined in Chapter 13)."""
    timer, num_batches = d2l.Timer(), len(train_iter)
    animator = d2l.Animator(xlabel='epoch', xlim=[1, num_epochs], ylim=[0, 1],
                            legend=['train loss', 'train acc', 'test acc'])
    net = nn.DataParallel(net, device_ids=devices).to(devices[0])
    for epoch in range(num_epochs):
        # Sum of training loss, sum of training accuracy, no. of examples,
        # no. of predictions
        metric = d2l.Accumulator(4)
        for i, (features, labels) in enumerate(train_iter):
            timer.start()
```

```
            l, acc = train_batch_ch13(
                net, features, labels, loss, trainer, devices)
            metric.add(l, acc, labels.shape[0], labels.numel())
            timer.stop()
            if (i + 1) % (num_batches // 5) == 0 or i == num_batches - 1:
                animator.add(epoch + (i + 1) / num_batches,
                             (metric[0] / metric[2], metric[1] / metric[3],
                              None))
        test_acc = d2l.evaluate_accuracy_gpu(net, test_iter)
        animator.add(epoch + 1, (None, None, test_acc))
    print(f'loss {metric[0] / metric[2]:.3f}, train acc '
          f'{metric[1] / metric[3]:.3f}, test acc {test_acc:.3f}')
    print(f'{metric[2] * num_epochs / timer.sum():.1f} examples/sec on '
          f'{str(devices)}')
```

Now we can define the `train_with_data_aug` function to train the model with image augmentation. This function gets all available GPUs, uses Adam as the optimization algorithm, applies image augmentation to the training dataset, and finally calls the `train_ch13` function just defined to train and evaluate the model.

```
batch_size, devices, net = 256, d2l.try_all_gpus(), d2l.resnet18(10, 3)
net.apply(d2l.init_cnn)

def train_with_data_aug(train_augs, test_augs, net, lr=0.001):
    train_iter = load_cifar10(True, train_augs, batch_size)
    test_iter = load_cifar10(False, test_augs, batch_size)
    loss = nn.CrossEntropyLoss(reduction="none")
    trainer = torch.optim.Adam(net.parameters(), lr=lr)
    net(next(iter(train_iter))[0])
    train_ch13(net, train_iter, test_iter, loss, trainer, 10, devices)
```

Let's train the model using image augmentation based on random left-right flipping.

```
train_with_data_aug(train_augs, test_augs, net)
```

```
loss 0.226, train acc 0.920, test acc 0.840
3322.3 examples/sec on [device(type='cuda', index=0), device(type='cuda',␣
↪index=1)]
```

### 14.1.3 Summary

- Image augmentation generates random images based on existing training data to improve the generalization ability of models.
- In order to obtain definitive results during prediction, we usually only apply image augmentation to training examples, and do not use image augmentation with random operations during prediction.

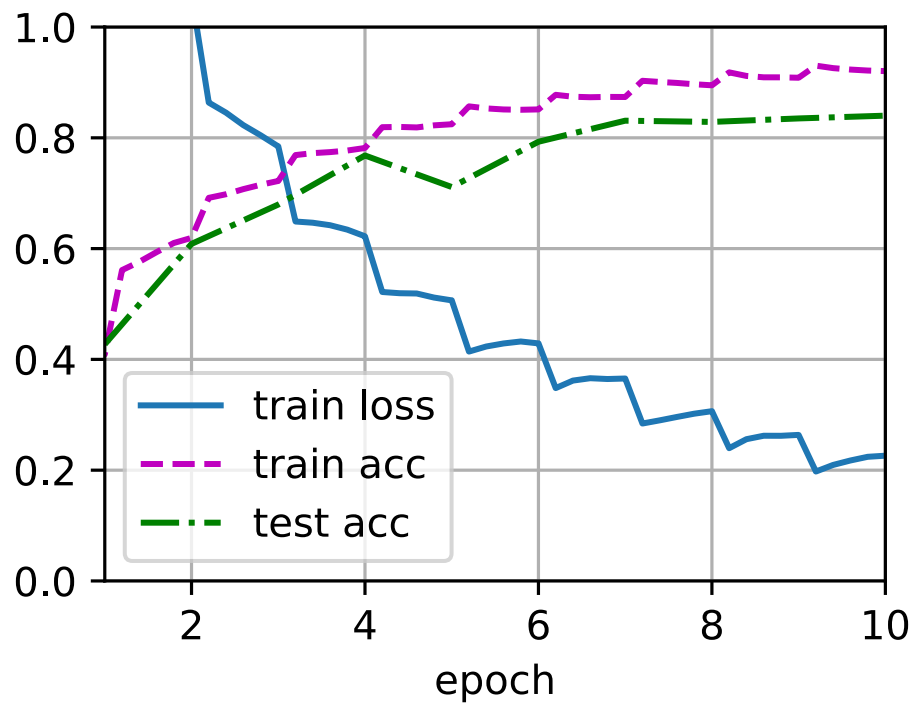

- Deep learning frameworks provide many different image augmentation methods, which can be applied simultaneously.

### 14.1.4 Exercises

1. Train the model without using image augmentation: `train_with_data_aug(test_augs, test_augs)`. Compare training and testing accuracy when using and not using image augmentation. Can this comparative experiment support the argument that image augmentation can mitigate overfitting? Why?
2. Combine multiple different image augmentation methods in model training on the CIFAR-10 dataset. Does it improve test accuracy?
3. Refer to the online documentation of the deep learning framework. What other image augmentation methods does it also provide?

Discussions[212].

## 14.2 Fine-Tuning

In earlier chapters, we discussed how to train models on the Fashion-MNIST training dataset with only 60000 images. We also described ImageNet, the most widely used large-scale image dataset in academia, which has more than 10 million images and 1000 objects. However, the size of the dataset that we usually encounter is between those of the two datasets.

Suppose that we want to recognize different types of chairs from images, and then recommend purchase links to users. One possible method is to first identify 100 common chairs, take 1000 images of different angles for each chair, and then train a classification model on the collected image dataset. Although this chair dataset may be larger than the Fashion-MNIST dataset, the number of examples is still less than one-tenth of that in ImageNet. This may lead to overfitting of complicated models that are suitable for ImageNet on this chair dataset. Besides, due to the limited amount of training examples, the accuracy of the trained model may not meet practical requirements.

In order to address the above problems, an obvious solution is to collect more data. However, collecting and labeling data can take a lot of time and money. For example, in order to collect the ImageNet dataset, researchers have spent millions of dollars from research funding. Although the current data collection cost has been significantly reduced, this cost still cannot be ignored.

Another solution is to apply *transfer learning* to transfer the knowledge learned from the *source dataset* to the *target dataset*. For example, although most of the images in the ImageNet dataset have nothing to do with chairs, the model trained on this dataset may extract more general image features, which can help identify edges, textures, shapes, and object composition. These similar features may also be effective for recognizing chairs.

### 14.2.1 Steps

In this section, we will introduce a common technique in transfer learning: *fine-tuning*. As shown in Fig. 14.2.1, fine-tuning consists of the following four steps:

1. Pretrain a neural network model, i.e., the *source model*, on a source dataset (e.g., the ImageNet dataset).
2. Create a new neural network model, i.e., the *target model*. This copies all model designs and their parameters on the source model except the output layer. We assume that these model parameters contain the knowledge learned from the source dataset and this knowledge will also be applicable to the target dataset. We also assume that the output layer of the source model is closely related to the labels of the source dataset; thus it is not used in the target model.
3. Add an output layer to the target model, whose number of outputs is the number of categories in the target dataset. Then randomly initialize the model parameters of this layer.
4. Train the target model on the target dataset, such as a chair dataset. The output layer will be trained from scratch, while the parameters of all the other layers are fine-tuned based on the parameters of the source model.

When target datasets are much smaller than source datasets, fine-tuning helps to improve models' generalization ability.

### 14.2.2 Hot Dog Recognition

Let's demonstrate fine-tuning via a concrete case: hot dog recognition. We will fine-tune a ResNet model on a small dataset, which was pretrained on the ImageNet dataset. This small dataset consists of thousands of images with and without hot dogs. We will use the fine-tuned model to recognize hot dogs from images.

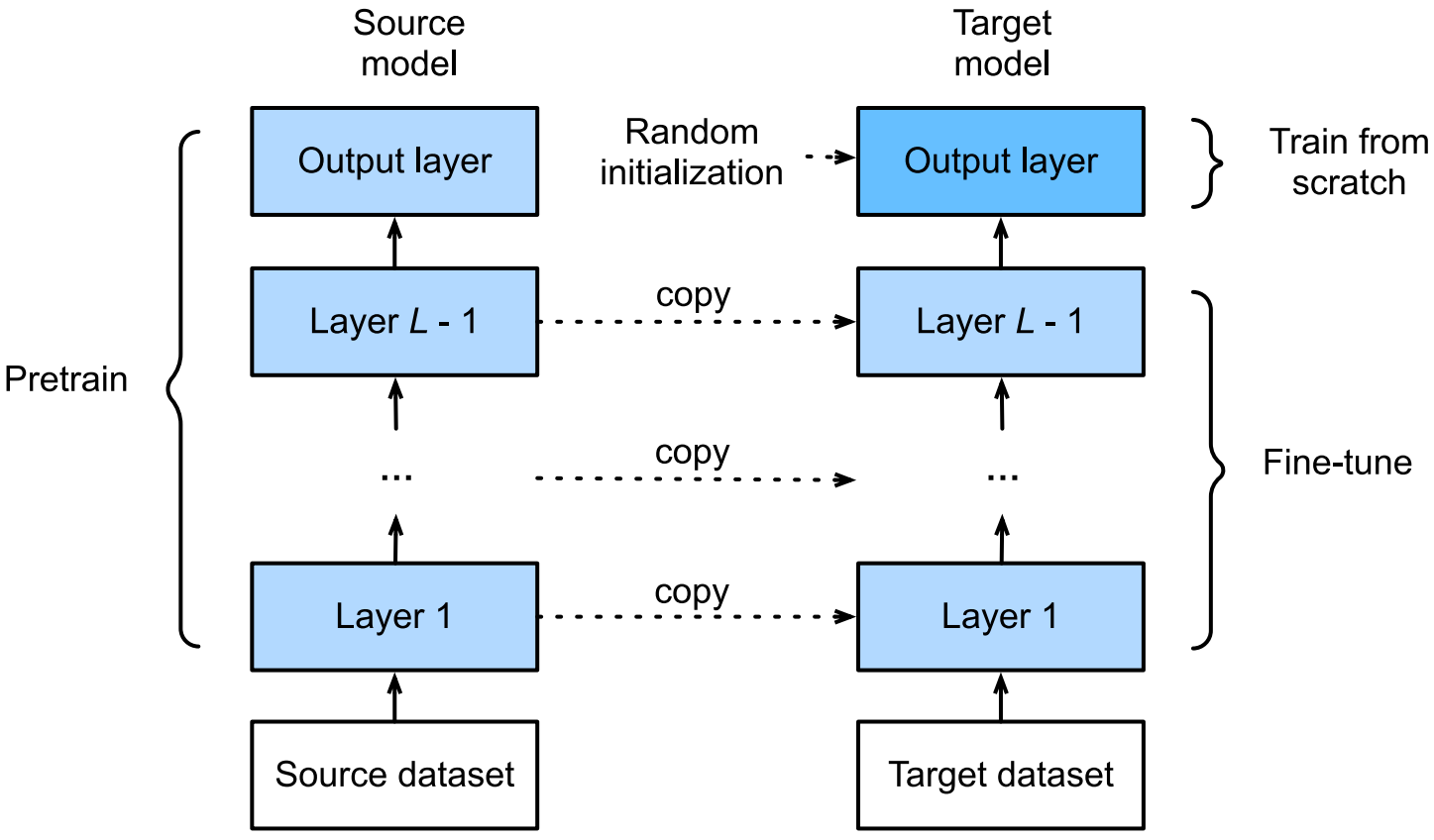

Fig. 14.2.1 Fine tuning.

```
%matplotlib inline
import os
import torch
import torchvision
from torch import nn
from d2l import torch as d2l
```

### Reading the Dataset

The hot dog dataset we use was taken from online images. This dataset consists of 1400 positive-class images containing hot dogs, and as many negative-class images containing other foods. 1000 images of both classes are used for training and the rest are for testing.

After unzipping the downloaded dataset, we obtain two folders `hotdog/train` and `hotdog/test`. Both folders have `hotdog` and `not-hotdog` subfolders, either of which contains images of the corresponding class.

```
#@save
d2l.DATA_HUB['hotdog'] = (d2l.DATA_URL + 'hotdog.zip',
                         'fba480ffa8aa7e0febbb511d181409f899b9baa5')

data_dir = d2l.download_extract('hotdog')
```

We create two instances to read all the image files in the training and testing datasets, respectively.

```
train_imgs = torchvision.datasets.ImageFolder(os.path.join(data_dir, 'train'))
test_imgs = torchvision.datasets.ImageFolder(os.path.join(data_dir, 'test'))
```

The first 8 positive examples and the last 8 negative images are shown below. As you can see, the images vary in size and aspect ratio.

```
hotdogs = [train_imgs[i][0] for i in range(8)]
not_hotdogs = [train_imgs[-i - 1][0] for i in range(8)]
d2l.show_images(hotdogs + not_hotdogs, 2, 8, scale=1.4);
```

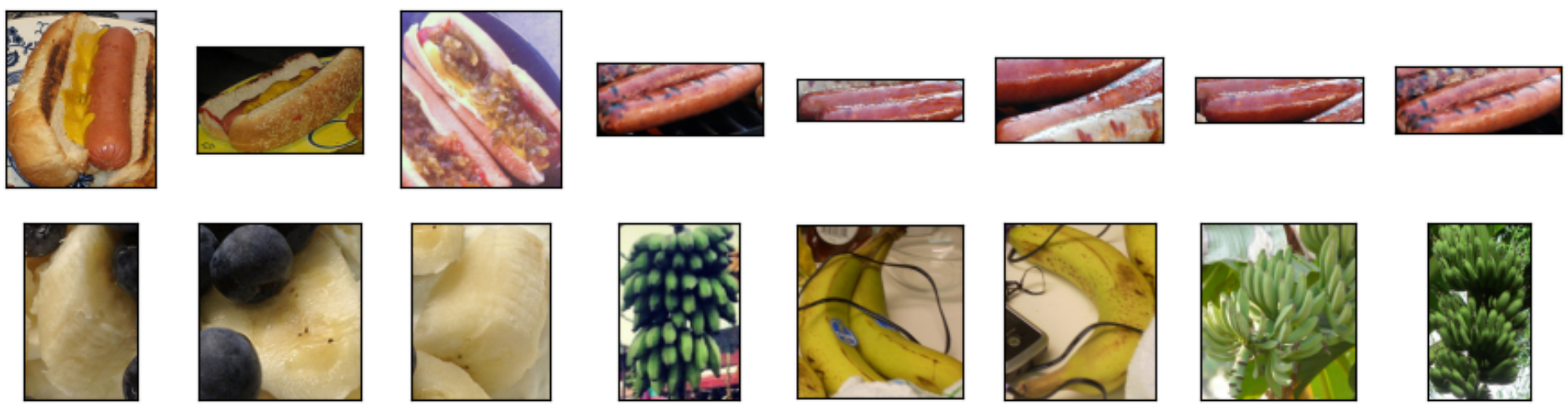

During training, we first crop a random area of random size and random aspect ratio from the image, and then scale this area to a $224 \times 224$ input image. During testing, we scale both the height and width of an image to 256 pixels, and then crop a central $224 \times 224$ area as input. In addition, for the three RGB (red, green, and blue) color channels we *standardize* their values channel by channel. Concretely, the mean value of a channel is subtracted from each value of that channel and then the result is divided by the standard deviation of that channel.

```
# Specify the means and standard deviations of the three RGB channels to
# standardize each channel
normalize = torchvision.transforms.Normalize(
    [0.485, 0.456, 0.406], [0.229, 0.224, 0.225])

train_augs = torchvision.transforms.Compose([
    torchvision.transforms.RandomResizedCrop(224),
    torchvision.transforms.RandomHorizontalFlip(),
    torchvision.transforms.ToTensor(),
    normalize])

test_augs = torchvision.transforms.Compose([
    torchvision.transforms.Resize([256, 256]),
    torchvision.transforms.CenterCrop(224),
    torchvision.transforms.ToTensor(),
    normalize])
```

### Defining and Initializing the Model

We use ResNet-18, which was pretrained on the ImageNet dataset, as the source model. Here, we specify `pretrained=True` to automatically download the pretrained model parameters. If this model is used for the first time, Internet connection is required for download.

```
pretrained_net = torchvision.models.resnet18(pretrained=True)
```

The pretrained source model instance contains a number of feature layers and an output layer `fc`. The main purpose of this division is to facilitate the fine-tuning of model parameters of all layers but the output layer. The member variable `fc` of source model is given below.

```
pretrained_net.fc
```

```
Linear(in_features=512, out_features=1000, bias=True)
```

As a fully connected layer, it transforms ResNet's final global average pooling outputs into 1000 class outputs of the ImageNet dataset. We then construct a new neural network as the target model. It is defined in the same way as the pretrained source model except that its number of outputs in the final layer is set to the number of classes in the target dataset (rather than 1000).

In the code below, the model parameters before the output layer of the target model instance `finetune_net` are initialized to model parameters of the corresponding layers from the source model. Since these model parameters were obtained via pretraining on ImageNet, they are effective. Therefore, we can only use a small learning rate to *fine-tune* such pretrained parameters. In contrast, model parameters in the output layer are randomly initialized and generally require a larger learning rate to be learned from scratch. Letting the base learning rate be $\eta$, a learning rate of $10\eta$ will be used to iterate the model parameters in the output layer.

```
finetune_net = torchvision.models.resnet18(pretrained=True)
finetune_net.fc = nn.Linear(finetune_net.fc.in_features, 2)
nn.init.xavier_uniform_(finetune_net.fc.weight);
```

### Fine-Tuning the Model

First, we define a training function `train_fine_tuning` that uses fine-tuning so it can be called multiple times.

```
# If `param_group=True`, the model parameters in the output layer will be
# updated using a learning rate ten times greater
def train_fine_tuning(net, learning_rate, batch_size=128, num_epochs=5,
                      param_group=True):
    train_iter = torch.utils.data.DataLoader(torchvision.datasets.ImageFolder(
        os.path.join(data_dir, 'train'), transform=train_augs),
        batch_size=batch_size, shuffle=True)
    test_iter = torch.utils.data.DataLoader(torchvision.datasets.ImageFolder(
        os.path.join(data_dir, 'test'), transform=test_augs),
        batch_size=batch_size)
```

```
    devices = d2l.try_all_gpus()
    loss = nn.CrossEntropyLoss(reduction="none")
    if param_group:
        params_1x = [param for name, param in net.named_parameters()
             if name not in ["fc.weight", "fc.bias"]]
        trainer = torch.optim.SGD([{'params': params_1x},
                                   {'params': net.fc.parameters(),
                                    'lr': learning_rate * 10}],
                                lr=learning_rate, weight_decay=0.001)
    else:
        trainer = torch.optim.SGD(net.parameters(), lr=learning_rate,
                                  weight_decay=0.001)
    d2l.train_ch13(net, train_iter, test_iter, loss, trainer, num_epochs,
                   devices)
```

We set the base learning rate to a small value in order to *fine-tune* the model parameters obtained via pretraining. Based on the previous settings, we will train the output layer parameters of the target model from scratch using a learning rate ten times greater.

```
train_fine_tuning(finetune_net, 5e-5)
```

```
loss 0.204, train acc 0.918, test acc 0.926
659.8 examples/sec on [device(type='cuda', index=0), device(type='cuda',␣
↪index=1)]
```

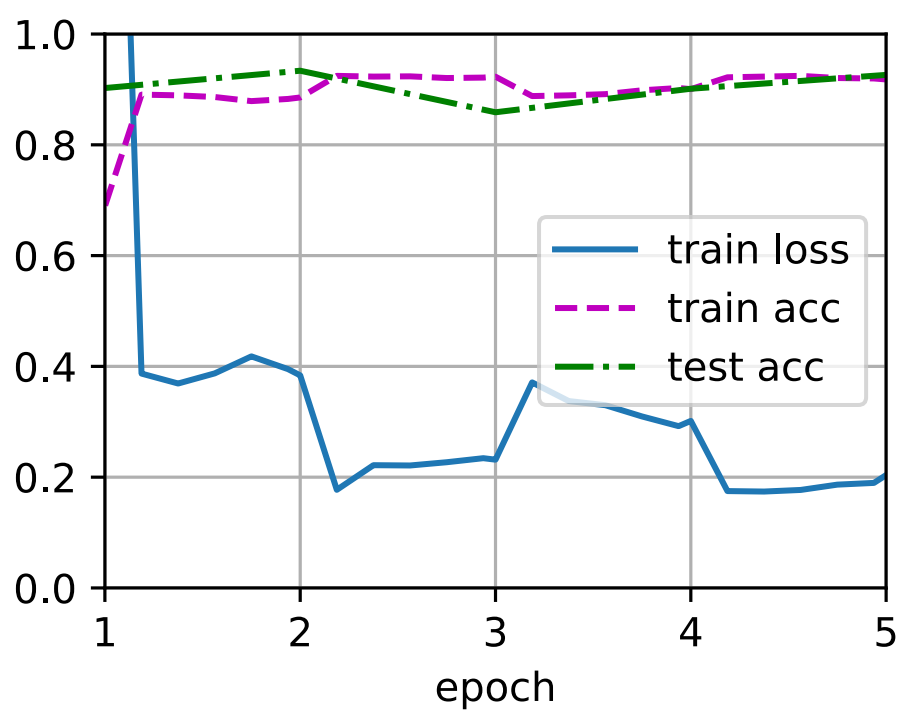

For comparison, we define an identical model, but initialize all of its model parameters to random values. Since the entire model needs to be trained from scratch, we can use a larger learning rate.

```
scratch_net = torchvision.models.resnet18()
scratch_net.fc = nn.Linear(scratch_net.fc.in_features, 2)
train_fine_tuning(scratch_net, 5e-4, param_group=False)
```

```
loss 0.459, train acc 0.819, test acc 0.871
1284.3 examples/sec on [device(type='cuda', index=0), device(type='cuda',␣
↪index=1)]
```

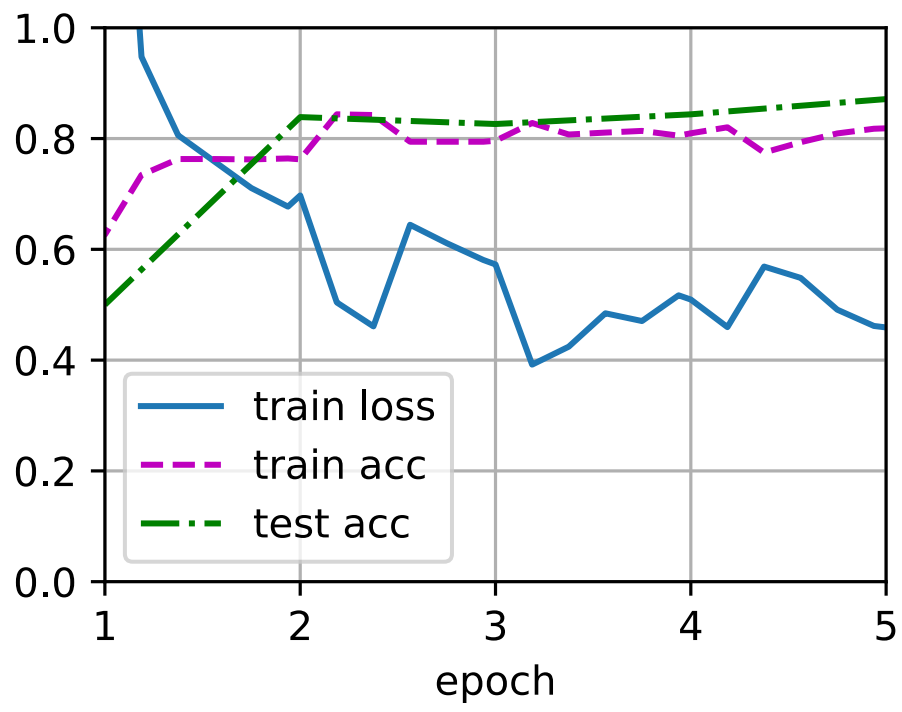

As we can see, the fine-tuned model tends to perform better for the same epoch because its initial parameter values are more effective.

## 14.2.3 Summary

- Transfer learning transfers knowledge learned from the source dataset to the target dataset. Fine-tuning is a common technique for transfer learning.
- The target model copies all model designs with their parameters from the source model except the output layer, and fine-tunes these parameters based on the target dataset. In contrast, the output layer of the target model needs to be trained from scratch.
- Generally, fine-tuning parameters uses a smaller learning rate, while training the output layer from scratch can use a larger learning rate.

## 14.2.4 Exercises

1. Keep increasing the learning rate of `finetune_net`. How does the accuracy of the model change?
2. Further adjust hyperparameters of `finetune_net` and `scratch_net` in the comparative experiment. Do they still differ in accuracy?
3. Set the parameters before the output layer of `finetune_net` to those of the source model and do *not* update them during training. How does the accuracy of the model change? You can use the following code.

```
for param in finetune_net.parameters():
    param.requires_grad = False
```

4. In fact, there is a "hotdog" class in the ImageNet dataset. Its corresponding weight parameter in the output layer can be obtained via the following code. How can we leverage this weight parameter?

```
weight = pretrained_net.fc.weight
hotdog_w = torch.split(weight.data, 1, dim=0)[934]
hotdog_w.shape
```

```
torch.Size([1, 512])
```

Discussions[213].

## 14.3 Object Detection and Bounding Boxes

In earlier sections (e.g., Section 8.1–Section 8.4), we introduced various models for image classification. In image classification tasks, we assume that there is only *one* major object in the image and we only focus on how to recognize its category. However, there are often *multiple* objects in the image of interest. We not only want to know their categories, but also their specific positions in the image. In computer vision, we refer to such tasks as *object detection* (or *object recognition*).

Object detection has been widely applied in many fields. For example, self-driving needs to plan traveling routes by detecting the positions of vehicles, pedestrians, roads, and obstacles in the captured video images. Besides, robots may use this technique to detect and localize objects of interest throughout its navigation of an environment. Moreover, security systems may need to detect abnormal objects, such as intruders or bombs.

In the next few sections, we will introduce several deep learning methods for object detection. We will begin with an introduction to *positions* (or *locations*) of objects.

```
%matplotlib inline
import torch
from d2l import torch as d2l
```

We will load the sample image to be used in this section. We can see that there is a dog on the left side of the image and a cat on the right. They are the two major objects in this image.

```
d2l.set_figsize()
img = d2l.plt.imread('../img/catdog.jpg')
d2l.plt.imshow(img);
```

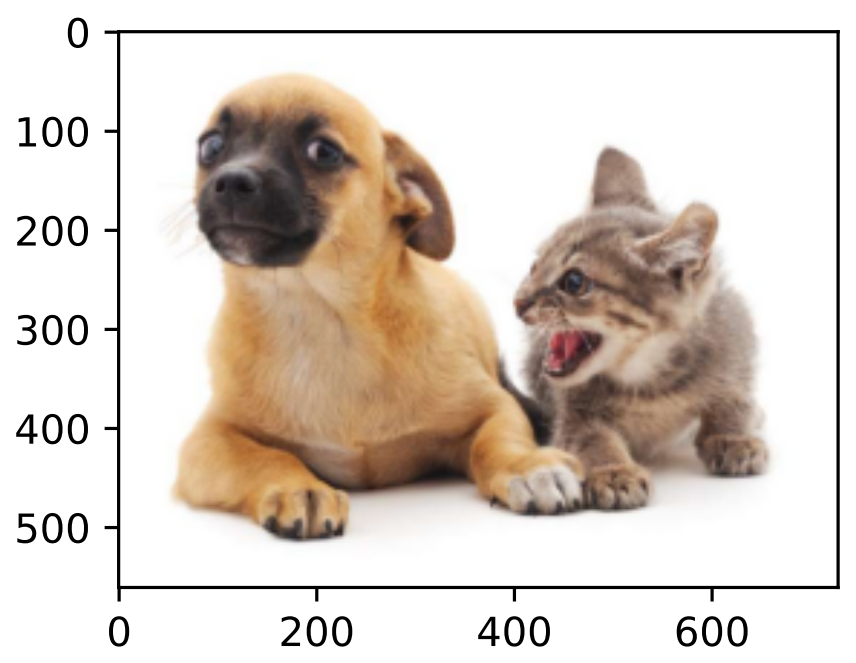

### 14.3.1 Bounding Boxes

In object detection, we usually use a *bounding box* to describe the spatial location of an object. The bounding box is rectangular, which is determined by the $x$ and $y$ coordinates of the upper-left corner of the rectangle and the such coordinates of the lower-right corner. Another commonly used bounding box representation is the $(x, y)$-axis coordinates of the bounding box center, and the width and height of the box.

Here we define functions to convert between these two representations: `box_corner_to_center` converts from the two-corner representation to the center-width-height presentation, and `box_center_to_corner` vice versa. The input argument `boxes` should be a two-dimensional tensor of shape $(n, 4)$, where $n$ is the number of bounding boxes.

```
#@save
def box_corner_to_center(boxes):
    """Convert from (upper-left, lower-right) to (center, width, height)."""
    x1, y1, x2, y2 = boxes[:, 0], boxes[:, 1], boxes[:, 2], boxes[:, 3]
    cx = (x1 + x2) / 2
    cy = (y1 + y2) / 2
    w = x2 - x1
    h = y2 - y1
    boxes = torch.stack((cx, cy, w, h), axis=-1)
    return boxes

#@save
def box_center_to_corner(boxes):
    """Convert from (center, width, height) to (upper-left, lower-right)."""
    cx, cy, w, h = boxes[:, 0], boxes[:, 1], boxes[:, 2], boxes[:, 3]
    x1 = cx - 0.5 * w
    y1 = cy - 0.5 * h
    x2 = cx + 0.5 * w
    y2 = cy + 0.5 * h
    boxes = torch.stack((x1, y1, x2, y2), axis=-1)
    return boxes
```

We will define the bounding boxes of the dog and the cat in the image based on the coordinate information. The origin of the coordinates in the image is the upper-left corner of the image, and to the right and down are the positive directions of the $x$ and $y$ axes, respectively.

```
# Here `bbox` is the abbreviation for bounding box
dog_bbox, cat_bbox = [60.0, 45.0, 378.0, 516.0], [400.0, 112.0, 655.0, 493.0]
```

We can verify the correctness of the two bounding box conversion functions by converting twice.

```
boxes = torch.tensor((dog_bbox, cat_bbox))
box_center_to_corner(box_corner_to_center(boxes)) == boxes
```

```
tensor([[True, True, True, True],
        [True, True, True, True]])
```

Let's draw the bounding boxes in the image to check if they are accurate. Before drawing, we will define a helper function `bbox_to_rect`. It represents the bounding box in the bounding box format of the `matplotlib` package.

```
#@save
def bbox_to_rect(bbox, color):
    """Convert bounding box to matplotlib format."""
    # Convert the bounding box (upper-left x, upper-left y, lower-right x,
    # lower-right y) format to the matplotlib format: ((upper-left x,
    # upper-left y), width, height)
    return d2l.plt.Rectangle(
        xy=(bbox[0], bbox[1]), width=bbox[2]-bbox[0], height=bbox[3]-bbox[1],
        fill=False, edgecolor=color, linewidth=2)
```

After adding the bounding boxes on the image, we can see that the main outline of the two objects are basically inside the two boxes.

```
fig = d2l.plt.imshow(img)
fig.axes.add_patch(bbox_to_rect(dog_bbox, 'blue'))
fig.axes.add_patch(bbox_to_rect(cat_bbox, 'red'));
```

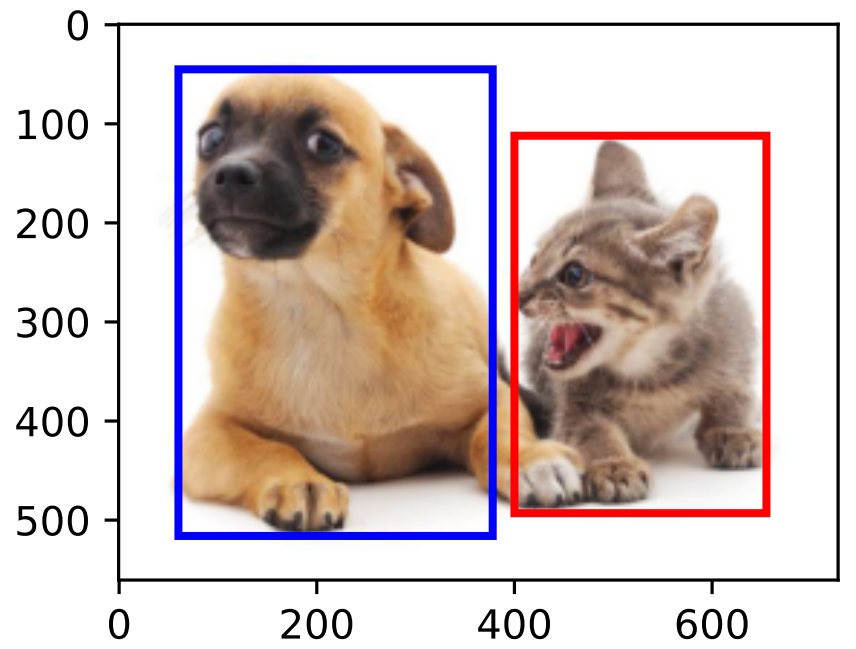

## 14.3.2 Summary

- Object detection not only recognizes all the objects of interest in the image, but also their positions. The position is generally represented by a rectangular bounding box.
- We can convert between two commonly used bounding box representations.

### 14.3.3 Exercises

1. Find another image and try to label a bounding box that contains the object. Compare labeling bounding boxes and categories: which usually takes longer?
2. Why is the innermost dimension of the input argument `boxes` of `box_corner_to_center` and `box_center_to_corner` always 4?

Discussions[214].

## 14.4 Anchor Boxes

Object detection algorithms usually sample a large number of regions in the input image, determine whether these regions contain objects of interest, and adjust the boundaries of the regions so as to predict the *ground-truth bounding boxes* of the objects more accurately. Different models may adopt different region sampling schemes. Here we introduce one of such methods: it generates multiple bounding boxes with varying scales and aspect ratios centered on each pixel. These bounding boxes are called *anchor boxes*. We will design an object detection model based on anchor boxes in Section 14.7.

First, let's modify the printing accuracy just for more concise outputs.

```
%matplotlib inline
import torch
from d2l import torch as d2l

torch.set_printoptions(2)  # Simplify printing accuracy
```

### 14.4.1 Generating Multiple Anchor Boxes

Suppose that the input image has a height of $h$ and width of $w$. We generate anchor boxes with different shapes centered on each pixel of the image. Let the *scale* be $s \in (0, 1]$ and the *aspect ratio* (ratio of width to height) is $r > 0$. Then the width and height of the anchor box are $ws\sqrt{r}$ and $hs/\sqrt{r}$, respectively. Note that when the center position is given, an anchor box with known width and height is determined.

To generate multiple anchor boxes with different shapes, let's set a series of scales $s_1, \ldots, s_n$

and a series of aspect ratios $r_1, \ldots, r_m$. When using all the combinations of these scales and aspect ratios with each pixel as the center, the input image will have a total of $whnm$ anchor boxes. Although these anchor boxes may cover all the ground-truth bounding boxes, the computational complexity is easily too high. In practice, we can only consider those combinations containing $s_1$ or $r_1$:

$$(s_1, r_1), (s_1, r_2), \ldots, (s_1, r_m), (s_2, r_1), (s_3, r_1), \ldots, (s_n, r_1). \tag{14.4.1}$$

That is to say, the number of anchor boxes centered on the same pixel is $n + m - 1$. For the entire input image, we will generate a total of $wh(n + m - 1)$ anchor boxes.

The above method of generating anchor boxes is implemented in the following `multibox_prior` function. We specify the input image, a list of scales, and a list of aspect ratios, then this function will return all the anchor boxes.

```
#@save
def multibox_prior(data, sizes, ratios):
    """Generate anchor boxes with different shapes centered on each pixel."""
    in_height, in_width = data.shape[-2:]
    device, num_sizes, num_ratios = data.device, len(sizes), len(ratios)
    boxes_per_pixel = (num_sizes + num_ratios - 1)
    size_tensor = torch.tensor(sizes, device=device)
    ratio_tensor = torch.tensor(ratios, device=device)
    # Offsets are required to move the anchor to the center of a pixel. Since
    # a pixel has height=1 and width=1, we choose to offset our centers by 0.5
    offset_h, offset_w = 0.5, 0.5
    steps_h = 1.0 / in_height  # Scaled steps in y axis
    steps_w = 1.0 / in_width  # Scaled steps in x axis

    # Generate all center points for the anchor boxes
    center_h = (torch.arange(in_height, device=device) + offset_h) * steps_h
    center_w = (torch.arange(in_width, device=device) + offset_w) * steps_w
    shift_y, shift_x = torch.meshgrid(center_h, center_w, indexing='ij')
    shift_y, shift_x = shift_y.reshape(-1), shift_x.reshape(-1)

    # Generate `boxes_per_pixel` number of heights and widths that are later
    # used to create anchor box corner coordinates (xmin, xmax, ymin, ymax)
    w = torch.cat((size_tensor * torch.sqrt(ratio_tensor[0]),
                   sizes[0] * torch.sqrt(ratio_tensor[1:])))\
                   * in_height / in_width  # Handle rectangular inputs
    h = torch.cat((size_tensor / torch.sqrt(ratio_tensor[0]),
                   sizes[0] / torch.sqrt(ratio_tensor[1:])))
    # Divide by 2 to get half height and half width
    anchor_manipulations = torch.stack((-w, -h, w, h)).T.repeat(
                                        in_height * in_width, 1) / 2

    # Each center point will have `boxes_per_pixel` number of anchor boxes, so
    # generate a grid of all anchor box centers with `boxes_per_pixel` repeats
    out_grid = torch.stack([shift_x, shift_y, shift_x, shift_y],
                dim=1).repeat_interleave(boxes_per_pixel, dim=0)
    output = out_grid + anchor_manipulations
    return output.unsqueeze(0)
```

We can see that the shape of the returned anchor box variable `Y` is (batch size, number of anchor boxes, 4).

```
img = d2l.plt.imread('../img/catdog.jpg')
h, w = img.shape[:2]

print(h, w)
X = torch.rand(size=(1, 3, h, w))  # Construct input data
Y = multibox_prior(X, sizes=[0.75, 0.5, 0.25], ratios=[1, 2, 0.5])
Y.shape
```

```
561 728
```

```
torch.Size([1, 2042040, 4])
```

After changing the shape of the anchor box variable `Y` to (image height, image width, number of anchor boxes centered on the same pixel, 4), we can obtain all the anchor boxes centered on a specified pixel position. In the following, we access the first anchor box centered on (250, 250). It has four elements: the $(x, y)$-axis coordinates at the upper-left corner and the $(x, y)$-axis coordinates at the lower-right corner of the anchor box. The coordinate values of both axes are divided by the width and height of the image, respectively.

```
boxes = Y.reshape(h, w, 5, 4)
boxes[250, 250, 0, :]
```

```
tensor([0.06, 0.07, 0.63, 0.82])
```

In order to show all the anchor boxes centered on one pixel in the image, we define the following `show_bboxes` function to draw multiple bounding boxes on the image.

```
#@save
def show_bboxes(axes, bboxes, labels=None, colors=None):
    """Show bounding boxes."""

    def make_list(obj, default_values=None):
        if obj is None:
            obj = default_values
        elif not isinstance(obj, (list, tuple)):
            obj = [obj]
        return obj

    labels = make_list(labels)
    colors = make_list(colors, ['b', 'g', 'r', 'm', 'c'])
    for i, bbox in enumerate(bboxes):
        color = colors[i % len(colors)]
        rect = d2l.bbox_to_rect(bbox.detach().numpy(), color)
        axes.add_patch(rect)
        if labels and len(labels) > i:
```

```
            text_color = 'k' if color == 'w' else 'w'
            axes.text(rect.xy[0], rect.xy[1], labels[i],
                      va='center', ha='center', fontsize=9, color=text_color,
                      bbox=dict(facecolor=color, lw=0))
```

As we just saw, the coordinate values of the $x$ and $y$ axes in the variable `boxes` have been divided by the width and height of the image, respectively. When drawing anchor boxes, we need to restore their original coordinate values; thus, we define variable `bbox_scale` below. Now, we can draw all the anchor boxes centered on (250, 250) in the image. As you can see, the blue anchor box with a scale of 0.75 and an aspect ratio of 1 well surrounds the dog in the image.

```
d2l.set_figsize()
bbox_scale = torch.tensor((w, h, w, h))
fig = d2l.plt.imshow(img)
show_bboxes(fig.axes, boxes[250, 250, :, :] * bbox_scale,
            ['s=0.75, r=1', 's=0.5, r=1', 's=0.25, r=1', 's=0.75, r=2',
             's=0.75, r=0.5'])
```

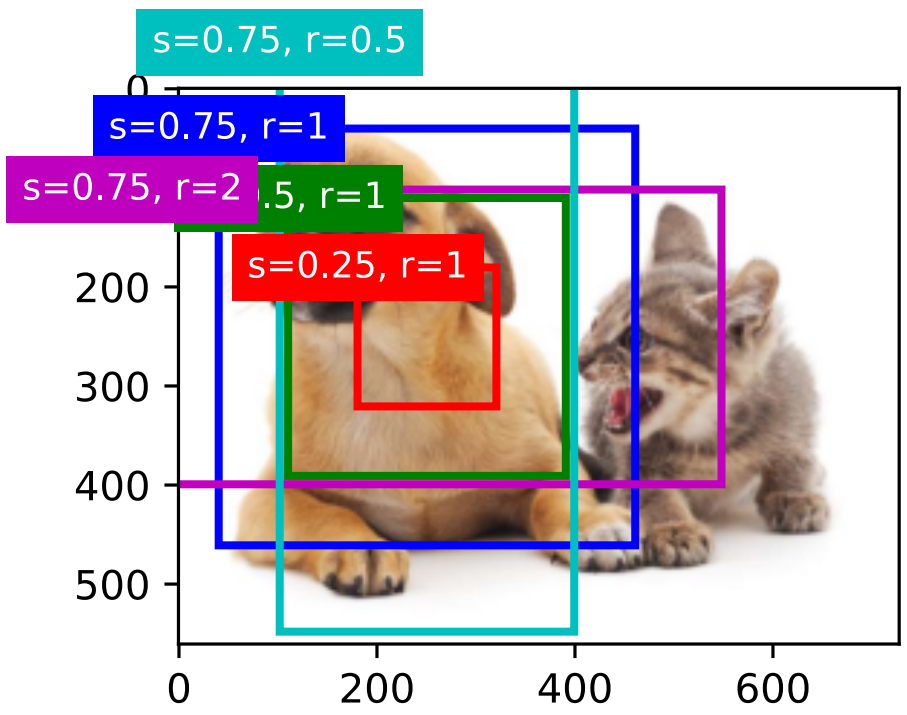

## 14.4.2 Intersection over Union (IoU)

We just mentioned that an anchor box "well" surrounds the dog in the image. If the ground-truth bounding box of the object is known, how can "well" here be quantified? Intuitively, we can measure the similarity between the anchor box and the ground-truth bounding box. We know that the *Jaccard index* can measure the similarity between two sets. Given sets $\mathcal{A}$ and $\mathcal{B}$, their Jaccard index is the size of their intersection divided by the size of their union:

$$J(\mathcal{A},\mathcal{B}) = \frac{\left|\mathcal{A} \cap \mathcal{B}\right|}{\left| \mathcal{A} \cup \mathcal{B}\right|}. \tag{14.4.2}$$

In fact, we can consider the pixel area of any bounding box as a set of pixels. In this way, we can measure the similarity of the two bounding boxes by the Jaccard index of their pixel sets. For two bounding boxes, we usually refer their Jaccard index as *intersection over union* (*IoU*), which is the ratio of their intersection area to their union area, as shown in Fig. 14.4.1. The range of an IoU is between 0 and 1: 0 means that two bounding boxes do not overlap at all, while 1 indicates that the two bounding boxes are equal.

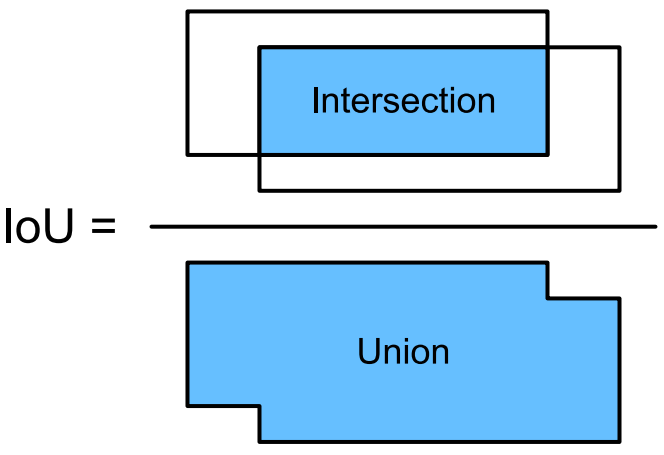

Fig. 14.4.1 IoU is the ratio of the intersection area to the union area of two bounding boxes.

For the remainder of this section, we will use IoU to measure the similarity between anchor boxes and ground-truth bounding boxes, and between different anchor boxes. Given two lists of anchor or bounding boxes, the following `box_iou` computes their pairwise IoU across these two lists.

```
#@save
def box_iou(boxes1, boxes2):
    """Compute pairwise IoU across two lists of anchor or bounding boxes."""
    box_area = lambda boxes: ((boxes[:, 2] - boxes[:, 0]) *
                              (boxes[:, 3] - boxes[:, 1]))
    # Shape of `boxes1`, `boxes2`, `areas1`, `areas2`: (no. of boxes1, 4),
    # (no. of boxes2, 4), (no. of boxes1,), (no. of boxes2,)
    areas1 = box_area(boxes1)
    areas2 = box_area(boxes2)
    # Shape of `inter_upperlefts`, `inter_lowerrights`, `inters`: (no. of
    # boxes1, no. of boxes2, 2)
    inter_upperlefts = torch.max(boxes1[:, None, :2], boxes2[:, :2])
    inter_lowerrights = torch.min(boxes1[:, None, 2:], boxes2[:, 2:])
    inters = (inter_lowerrights - inter_upperlefts).clamp(min=0)
    # Shape of `inter_areas` and `union_areas`: (no. of boxes1, no. of boxes2)
    inter_areas = inters[:, :, 0] * inters[:, :, 1]
    union_areas = areas1[:, None] + areas2 - inter_areas
    return inter_areas / union_areas
```

## 14.4.3 Labeling Anchor Boxes in Training Data

In a training dataset, we consider each anchor box as a training example. In order to train an object detection model, we need *class* and *offset* labels for each anchor box, where the former is the class of the object relevant to the anchor box and the latter is the offset of the ground-truth bounding box relative to the anchor box. During the prediction, for each image we generate multiple anchor boxes, predict classes and offsets for all the anchor boxes, adjust their positions according to the predicted offsets to obtain the predicted bounding boxes, and finally only output those predicted bounding boxes that satisfy certain criteria.

As we know, an object detection training set comes with labels for locations of *ground-truth*

*bounding boxes* and classes of their surrounded objects. To label any generated *anchor box*, we refer to the labeled location and class of its *assigned* ground-truth bounding box that is closest to the anchor box. In the following, we describe an algorithm for assigning closest ground-truth bounding boxes to anchor boxes.

### Assigning Ground-Truth Bounding Boxes to Anchor Boxes

Given an image, suppose that the anchor boxes are $A_1, A_2, \ldots, A_{n_a}$ and the ground-truth bounding boxes are $B_1, B_2, \ldots, B_{n_b}$, where $n_a \geq n_b$. Let's define a matrix $\mathbf{X} \in \mathbb{R}^{n_a \times n_b}$, whose element $x_{ij}$ in the $i^\textrm{th}$ row and $j^\textrm{th}$ column is the IoU of the anchor box $A_i$ and the ground-truth bounding box $B_j$. The algorithm consists of the following steps:

1. Find the largest element in matrix $\mathbf{X}$ and denote its row and column indices as $i_1$ and $j_1$, respectively. Then the ground-truth bounding box $B_{j_1}$ is assigned to the anchor box $A_{i_1}$. This is quite intuitive because $A_{i_1}$ and $B_{j_1}$ are the closest among all the pairs of anchor boxes and ground-truth bounding boxes. After the first assignment, discard all the elements in the ${i_1}^\mathrm{th}$ row and the ${j_1}^\mathrm{th}$ column in matrix $\mathbf{X}$.

2. Find the largest of the remaining elements in matrix $\mathbf{X}$ and denote its row and column indices as $i_2$ and $j_2$, respectively. We assign ground-truth bounding box $B_{j_2}$ to anchor box $A_{i_2}$ and discard all the elements in the ${i_2}^\mathrm{th}$ row and the ${j_2}^\mathrm{th}$ column in matrix $\mathbf{X}$.

3. At this point, elements in two rows and two columns in matrix $\mathbf{X}$ have been discarded. We proceed until all elements in $n_b$ columns in matrix $\mathbf{X}$ are discarded. At this time, we have assigned a ground-truth bounding box to each of $n_b$ anchor boxes.

4. Only traverse through the remaining $n_a - n_b$ anchor boxes. For example, given any anchor box $A_i$, find the ground-truth bounding box $B_j$ with the largest IoU with $A_i$ throughout the $i^\mathrm{th}$ row of matrix $\mathbf{X}$, and assign $B_j$ to $A_i$ only if this IoU is greater than a predefined threshold.

Let's illustrate the above algorithm using a concrete example. As shown in Fig. 14.4.2 (left), assuming that the maximum value in matrix $\mathbf{X}$ is $x_{23}$, we assign the ground-truth bounding box $B_3$ to the anchor box $A_2$. Then, we discard all the elements in row 2 and column 3 of the matrix, find the largest $x_{71}$ in the remaining elements (shaded area), and assign the ground-truth bounding box $B_1$ to the anchor box $A_7$. Next, as shown in Fig. 14.4.2 (middle), discard all the elements in row 7 and column 1 of the matrix, find the largest $x_{54}$ in the remaining elements (shaded area), and assign the ground-truth bounding box $B_4$ to the anchor box $A_5$. Finally, as shown in Fig. 14.4.2 (right), discard all the elements in row 5 and column 4 of the matrix, find the largest $x_{92}$ in the remaining elements (shaded area), and assign the ground-truth bounding box $B_2$ to the anchor box $A_9$. After that, we only need to traverse through the remaining anchor boxes $A_1, A_3, A_4, A_6, A_8$ and determine whether to assign them ground-truth bounding boxes according to the threshold.

This algorithm is implemented in the following `assign_anchor_to_bbox` function.

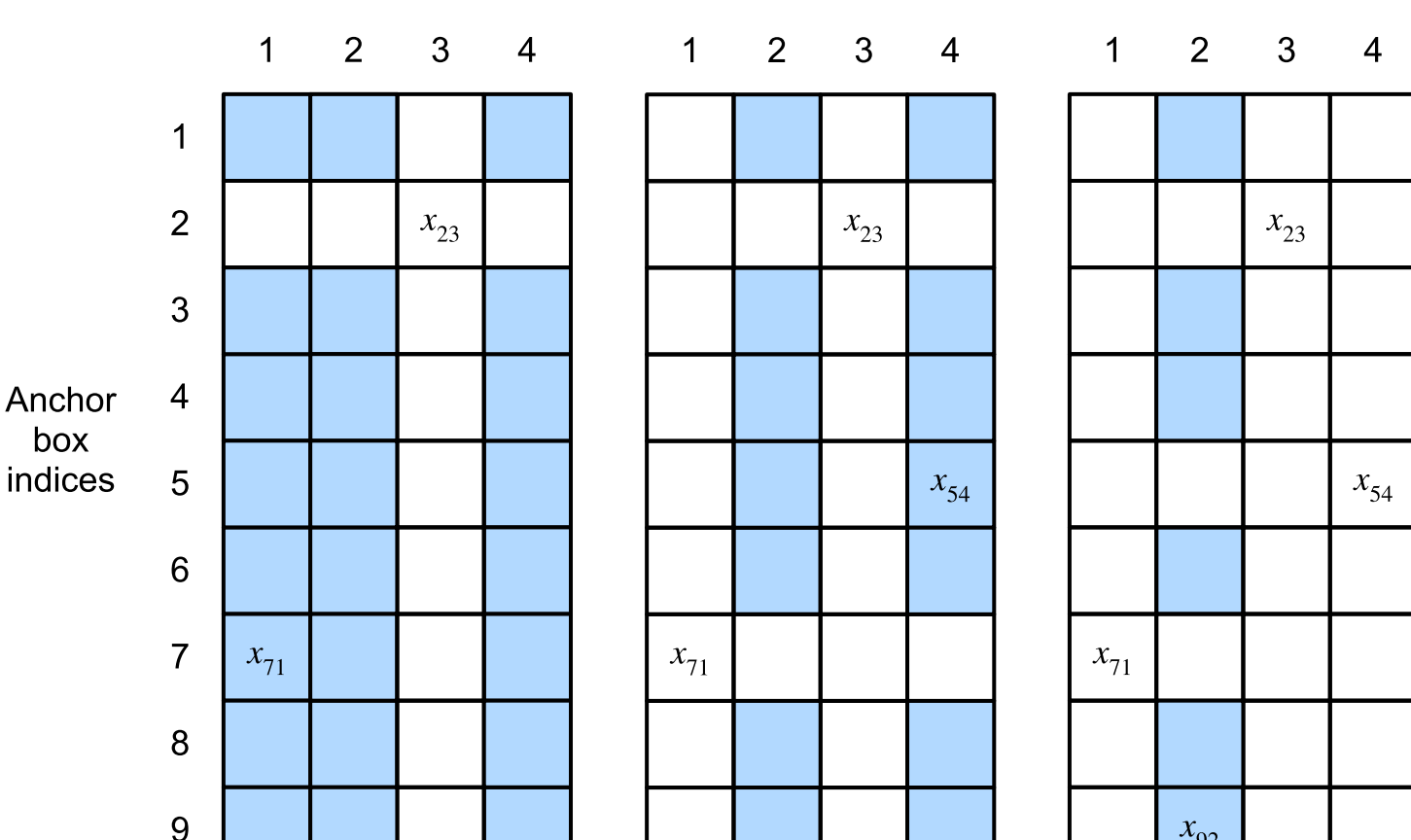

Fig. 14.4.2 Assigning ground-truth bounding boxes to anchor boxes.

```
#@save
def assign_anchor_to_bbox(ground_truth, anchors, device, iou_threshold=0.5):
    """Assign closest ground-truth bounding boxes to anchor boxes."""
    num_anchors, num_gt_boxes = anchors.shape[0], ground_truth.shape[0]
    # Element x_ij in the i-th row and j-th column is the IoU of the anchor
    # box i and the ground-truth bounding box j
    jaccard = box_iou(anchors, ground_truth)
    # Initialize the tensor to hold the assigned ground-truth bounding box for
    # each anchor
    anchors_bbox_map = torch.full((num_anchors,), -1, dtype=torch.long,
                                  device=device)
    # Assign ground-truth bounding boxes according to the threshold
    max_ious, indices = torch.max(jaccard, dim=1)
    anc_i = torch.nonzero(max_ious >= iou_threshold).reshape(-1)
    box_j = indices[max_ious >= iou_threshold]
    anchors_bbox_map[anc_i] = box_j
    col_discard = torch.full((num_anchors,), -1)
    row_discard = torch.full((num_gt_boxes,), -1)
    for _ in range(num_gt_boxes):
        max_idx = torch.argmax(jaccard)  # Find the largest IoU
        box_idx = (max_idx % num_gt_boxes).long()
        anc_idx = (max_idx / num_gt_boxes).long()
        anchors_bbox_map[anc_idx] = box_idx
        jaccard[:, box_idx] = col_discard
        jaccard[anc_idx, :] = row_discard
    return anchors_bbox_map
```

### Labeling Classes and Offsets

Now we can label the class and offset for each anchor box. Suppose that an anchor box $A$ is assigned a ground-truth bounding box $B$. On the one hand, the class of the anchor box $A$ will be labeled as that of $B$. On the other hand, the offset of the anchor box $A$ will be labeled according to the relative position between the central coordinates of $B$ and $A$ together with the relative size between these two boxes. Given varying positions and sizes of different boxes in the dataset, we can apply transformations to those relative positions and sizes that may lead to more uniformly distributed offsets that are easier to fit. Here we describe a common transformation. Given the central coordinates of $A$ and $B$ as $(x_a, y_a)$ and $(x_b, y_b)$, their widths as $w_a$ and $w_b$, and their heights as $h_a$ and $h_b$, respectively. We may label the offset of $A$ as

$$\left( \frac{\frac{x_b - x_a}{w_a} - \mu_x}{\sigma_x}, \frac{\frac{y_b - y_a}{h_a} - \mu_y}{\sigma_y}, \frac{\log \frac{w_b}{w_a} - \mu_w}{\sigma_w}, \frac{\log \frac{h_b}{h_a} - \mu_h}{\sigma_h} \right), \tag{14.4.3}$$

where default values of the constants are $\mu_x = \mu_y = \mu_w = \mu_h = 0, \sigma_x = \sigma_y = 0.1$, and $\sigma_w = \sigma_h = 0.2$. This transformation is implemented below in the `offset_boxes` function.

```
#@save
def offset_boxes(anchors, assigned_bb, eps=1e-6):
    """Transform for anchor box offsets."""
    c_anc = d2l.box_corner_to_center(anchors)
    c_assigned_bb = d2l.box_corner_to_center(assigned_bb)
    offset_xy = 10 * (c_assigned_bb[:, :2] - c_anc[:, :2]) / c_anc[:, 2:]
    offset_wh = 5 * torch.log(eps + c_assigned_bb[:, 2:] / c_anc[:, 2:])
    offset = torch.cat([offset_xy, offset_wh], axis=1)
    return offset
```

If an anchor box is not assigned a ground-truth bounding box, we just label the class of the anchor box as "background". Anchor boxes whose classes are background are often referred to as *negative* anchor boxes, and the rest are called *positive* anchor boxes. We implement the following `multibox_target` function to label classes and offsets for anchor boxes (the `anchors` argument) using ground-truth bounding boxes (the `labels` argument). This function sets the background class to zero and increments the integer index of a new class by one.

```
#@save
def multibox_target(anchors, labels):
    """Label anchor boxes using ground-truth bounding boxes."""
    batch_size, anchors = labels.shape[0], anchors.squeeze(0)
    batch_offset, batch_mask, batch_class_labels = [], [], []
    device, num_anchors = anchors.device, anchors.shape[0]
    for i in range(batch_size):
        label = labels[i, :, :]
        anchors_bbox_map = assign_anchor_to_bbox(
            label[:, 1:], anchors, device)
        bbox_mask = ((anchors_bbox_map >= 0).float().unsqueeze(-1)).repeat(
            1, 4)
        # Initialize class labels and assigned bounding box coordinates with
        # zeros
        class_labels = torch.zeros(num_anchors, dtype=torch.long,
                                   device=device)
```

```
        assigned_bb = torch.zeros((num_anchors, 4), dtype=torch.float32,
                                  device=device)
        # Label classes of anchor boxes using their assigned ground-truth
        # bounding boxes. If an anchor box is not assigned any, we label its
        # class as background (the value remains zero)
        indices_true = torch.nonzero(anchors_bbox_map >= 0)
        bb_idx = anchors_bbox_map[indices_true]
        class_labels[indices_true] = label[bb_idx, 0].long() + 1
        assigned_bb[indices_true] = label[bb_idx, 1:]
        # Offset transformation
        offset = offset_boxes(anchors, assigned_bb) * bbox_mask
        batch_offset.append(offset.reshape(-1))
        batch_mask.append(bbox_mask.reshape(-1))
        batch_class_labels.append(class_labels)
    bbox_offset = torch.stack(batch_offset)
    bbox_mask = torch.stack(batch_mask)
    class_labels = torch.stack(batch_class_labels)
    return (bbox_offset, bbox_mask, class_labels)
```

### An Example

Let's illustrate anchor box labeling via a concrete example. We define ground-truth bounding boxes for the dog and cat in the loaded image, where the first element is the class (0 for dog and 1 for cat) and the remaining four elements are the $(x, y)$-axis coordinates at the upper-left corner and the lower-right corner (range is between 0 and 1). We also construct five anchor boxes to be labeled using the coordinates of the upper-left corner and the lower-right corner: $A_0, \ldots, A_4$ (the index starts from 0). Then we plot these ground-truth bounding boxes and anchor boxes in the image.

```
ground_truth = torch.tensor([[0, 0.1, 0.08, 0.52, 0.92],
                             [1, 0.55, 0.2, 0.9, 0.88]])
anchors = torch.tensor([[0, 0.1, 0.2, 0.3], [0.15, 0.2, 0.4, 0.4],
                        [0.63, 0.05, 0.88, 0.98], [0.66, 0.45, 0.8, 0.8],
                        [0.57, 0.3, 0.92, 0.9]])

fig = d2l.plt.imshow(img)
show_bboxes(fig.axes, ground_truth[:, 1:] * bbox_scale, ['dog', 'cat'], 'k')
show_bboxes(fig.axes, anchors * bbox_scale, ['0', '1', '2', '3', '4']);
```

Using the `multibox_target` function defined above, we can label classes and offsets of these anchor boxes based on the ground-truth bounding boxes for the dog and cat. In this example, indices of the background, dog, and cat classes are 0, 1, and 2, respectively. Below we add an dimension for examples of anchor boxes and ground-truth bounding boxes.

```
labels = multibox_target(anchors.unsqueeze(dim=0),
                         ground_truth.unsqueeze(dim=0))
```

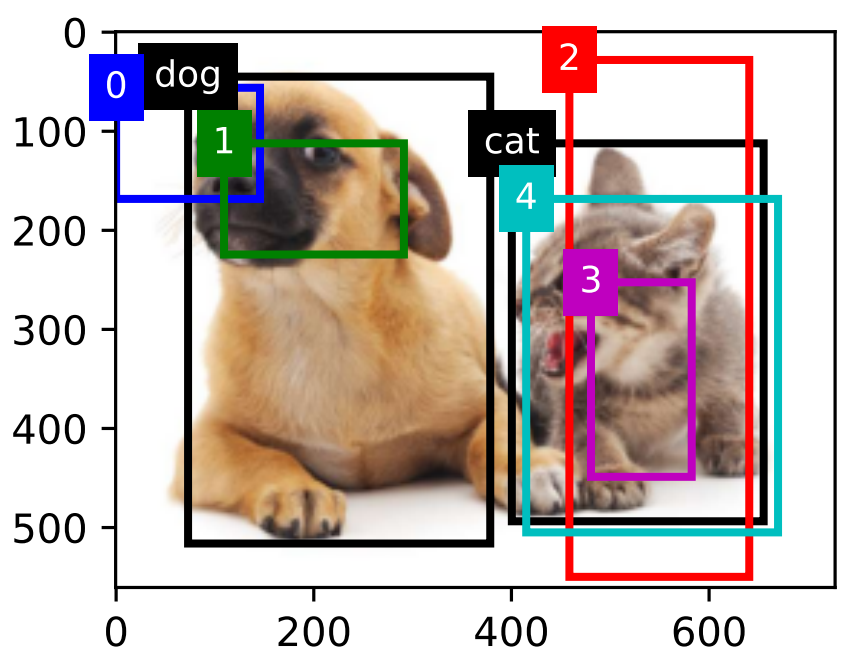

There are three items in the returned result, all of which are in the tensor format. The third item contains the labeled classes of the input anchor boxes.

Let's analyze the returned class labels below based on anchor box and ground-truth bounding box positions in the image. First, among all the pairs of anchor boxes and ground-truth bounding boxes, the IoU of the anchor box $A_4$ and the ground-truth bounding box of the cat is the largest. Thus, the class of $A_4$ is labeled as the cat. Taking out pairs containing $A_4$ or the ground-truth bounding box of the cat, among the rest the pair of the anchor box $A_1$ and the ground-truth bounding box of the dog has the largest IoU. So the class of $A_1$ is labeled as the dog. Next, we need to traverse through the remaining three unlabeled anchor boxes: $A_0$, $A_2$, and $A_3$. For $A_0$, the class of the ground-truth bounding box with the largest IoU is the dog, but the IoU is below the predefined threshold (0.5), so the class is labeled as background; for $A_2$, the class of the ground-truth bounding box with the largest IoU is the cat and the IoU exceeds the threshold, so the class is labeled as the cat; for $A_3$, the class of the ground-truth bounding box with the largest IoU is the cat, but the value is below the threshold, so the class is labeled as background.

```
labels[2]
```

```
tensor([[0, 1, 2, 0, 2]])
```

The second returned item is a mask variable of the shape (batch size, four times the number of anchor boxes). Every four elements in the mask variable correspond to the four offset values of each anchor box. Since we do not care about background detection, offsets of this negative class should not affect the objective function. Through elementwise multiplications, zeros in the mask variable will filter out negative class offsets before calculating the objective function.

```
labels[1]
```

```
tensor([[0., 0., 0., 0., 1., 1., 1., 1., 1., 1., 1., 1., 0., 0., 0., 0., 1., 1.
↪,
         1., 1.]])
```

The first returned item contains the four offset values labeled for each anchor box. Note that the offsets of negative-class anchor boxes are labeled as zeros.

```
labels[0]
```

```
tensor([[-0.00e+00, -0.00e+00, -0.00e+00, -0.00e+00,  1.40e+00,  1.00e+01,
          2.59e+00,  7.18e+00, -1.20e+00,  2.69e-01,  1.68e+00, -1.57e+00,
         -0.00e+00, -0.00e+00, -0.00e+00, -0.00e+00, -5.71e-01, -1.00e+00,
          4.17e-06,  6.26e-01]])
```

## 14.4.4 Predicting Bounding Boxes with Non-Maximum Suppression

During prediction, we generate multiple anchor boxes for the image and predict classes and offsets for each of them. A *predicted bounding box* is thus obtained according to an anchor box with its predicted offset. Below we implement the `offset_inverse` function that takes in anchors and offset predictions as inputs and applies inverse offset transformations to return the predicted bounding box coordinates.

```
#@save
def offset_inverse(anchors, offset_preds):
    """Predict bounding boxes based on anchor boxes with predicted offsets."""
    anc = d2l.box_corner_to_center(anchors)
    pred_bbox_xy = (offset_preds[:, :2] * anc[:, 2:] / 10) + anc[:, :2]
    pred_bbox_wh = torch.exp(offset_preds[:, 2:] / 5) * anc[:, 2:]
    pred_bbox = torch.cat((pred_bbox_xy, pred_bbox_wh), axis=1)
    predicted_bbox = d2l.box_center_to_corner(pred_bbox)
    return predicted_bbox
```

When there are many anchor boxes, many similar (with significant overlap) predicted bounding boxes can be potentially output for surrounding the same object. To simplify the output, we can merge similar predicted bounding boxes that belong to the same object by using *non-maximum suppression* (NMS).

Here is how non-maximum suppression works. For a predicted bounding box $B$, the object detection model calculates the predicted likelihood for each class. Denoting by $p$ the largest predicted likelihood, the class corresponding to this probability is the predicted class for $B$. Specifically, we refer to $p$ as the *confidence* (score) of the predicted bounding box $B$. On the same image, all the predicted non-background bounding boxes are sorted by confidence in descending order to generate a list $L$. Then we manipulate the sorted list $L$ in the following steps:

1. Select the predicted bounding box $B_1$ with the highest confidence from $L$ as a basis and remove all non-basis predicted bounding boxes whose IoU with $B_1$ exceeds a predefined threshold $\epsilon$ from $L$. At this point, $L$ keeps the predicted bounding box with the highest confidence but drops others that are too similar to it. In a nutshell, those with *non-maximum* confidence scores are *suppressed*.

2. Select the predicted bounding box $B_2$ with the second highest confidence from $L$ as another basis and remove all non-basis predicted bounding boxes whose IoU with $B_2$ exceeds $\epsilon$ from $L$.

3. Repeat the above process until all the predicted bounding boxes in $L$ have been used as a basis. At this time, the IoU of any pair of predicted bounding boxes in $L$ is below the threshold $\epsilon$; thus, no pair is too similar with each other.

4. Output all the predicted bounding boxes in the list $L$.

The following `nms` function sorts confidence scores in descending order and returns their indices.

```
#@save
def nms(boxes, scores, iou_threshold):
    """Sort confidence scores of predicted bounding boxes."""
    B = torch.argsort(scores, dim=-1, descending=True)
    keep = []  # Indices of predicted bounding boxes that will be kept
    while B.numel() > 0:
        i = B[0]
        keep.append(i)
        if B.numel() == 1: break
        iou = box_iou(boxes[i, :].reshape(-1, 4),
                      boxes[B[1:], :].reshape(-1, 4)).reshape(-1)
        inds = torch.nonzero(iou <= iou_threshold).reshape(-1)
        B = B[inds + 1]
    return torch.tensor(keep, device=boxes.device)
```

We define the following `multibox_detection` to apply non-maximum suppression to predicting bounding boxes. Do not worry if you find the implementation a bit complicated: we will show how it works with a concrete example right after the implementation.

```
#@save
def multibox_detection(cls_probs, offset_preds, anchors, nms_threshold=0.5,
                       pos_threshold=0.009999999):
    """Predict bounding boxes using non-maximum suppression."""
    device, batch_size = cls_probs.device, cls_probs.shape[0]
    anchors = anchors.squeeze(0)
    num_classes, num_anchors = cls_probs.shape[1], cls_probs.shape[2]
    out = []
    for i in range(batch_size):
        cls_prob, offset_pred = cls_probs[i], offset_preds[i].reshape(-1, 4)
        conf, class_id = torch.max(cls_prob[1:], 0)
        predicted_bb = offset_inverse(anchors, offset_pred)
        keep = nms(predicted_bb, conf, nms_threshold)
        # Find all non-`keep` indices and set the class to background
        all_idx = torch.arange(num_anchors, dtype=torch.long, device=device)
        combined = torch.cat((keep, all_idx))
        uniques, counts = combined.unique(return_counts=True)
        non_keep = uniques[counts == 1]
        all_id_sorted = torch.cat((keep, non_keep))
        class_id[non_keep] = -1
        class_id = class_id[all_id_sorted]
```

```
        conf, predicted_bb = conf[all_id_sorted], predicted_bb[all_id_sorted]
        # Here `pos_threshold` is a threshold for positive (non-background)
        # predictions
        below_min_idx = (conf < pos_threshold)
        class_id[below_min_idx] = -1
        conf[below_min_idx] = 1 - conf[below_min_idx]
        pred_info = torch.cat((class_id.unsqueeze(1),
                               conf.unsqueeze(1),
                               predicted_bb), dim=1)
        out.append(pred_info)
    return torch.stack(out)
```

Now let's apply the above implementations to a concrete example with four anchor boxes. For simplicity, we assume that the predicted offsets are all zeros. This means that the predicted bounding boxes are anchor boxes. For each class among the background, dog, and cat, we also define its predicted likelihood.

```
anchors = torch.tensor([[0.1, 0.08, 0.52, 0.92], [0.08, 0.2, 0.56, 0.95],
                      [0.15, 0.3, 0.62, 0.91], [0.55, 0.2, 0.9, 0.88]])
offset_preds = torch.tensor([0] * anchors.numel())
cls_probs = torch.tensor([[0] * 4,  # Predicted background likelihood
                      [0.9, 0.8, 0.7, 0.1],  # Predicted dog likelihood
                      [0.1, 0.2, 0.3, 0.9]])  # Predicted cat likelihood
```

We can plot these predicted bounding boxes with their confidence on the image.

```
fig = d2l.plt.imshow(img)
show_bboxes(fig.axes, anchors * bbox_scale,
            ['dog=0.9', 'dog=0.8', 'dog=0.7', 'cat=0.9'])
```

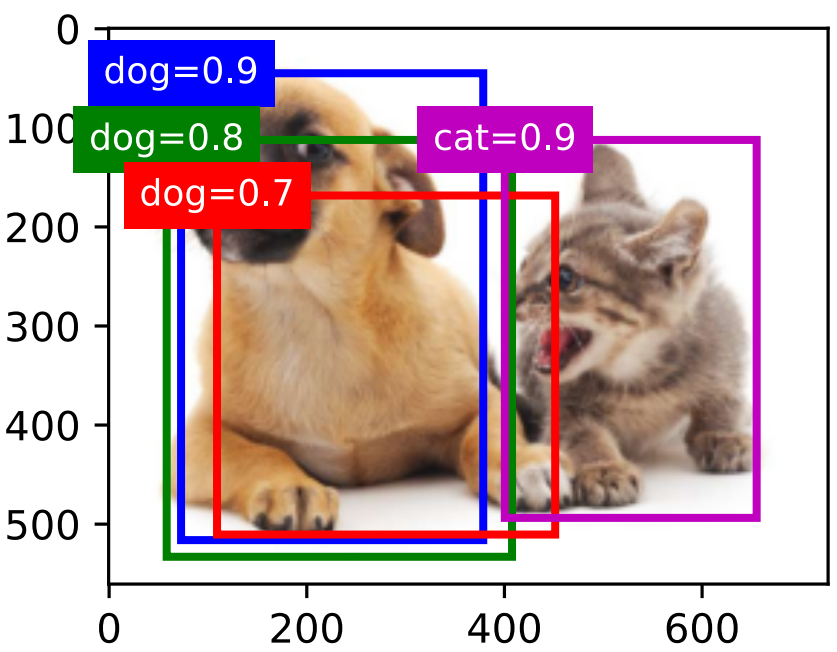

Now we can invoke the `multibox_detection` function to perform non-maximum suppression, where the threshold is set to 0.5. Note that we add a dimension for examples in the tensor input.

We can see that the shape of the returned result is (batch size, number of anchor boxes, 6). The six elements in the innermost dimension gives the output information for the same predicted bounding box. The first element is the predicted class index, which starts from 0 (0 is dog and 1 is cat). The value -1 indicates background or removal in non-maximum suppression. The second element is the confidence of the predicted bounding box. The remaining four elements are the $(x, y)$-axis coordinates of the upper-left corner and the lower-right corner of the predicted bounding box, respectively (range is between 0 and 1).

```
output = multibox_detection(cls_probs.unsqueeze(dim=0),
                            offset_preds.unsqueeze(dim=0),
                            anchors.unsqueeze(dim=0),
                            nms_threshold=0.5)
output
```

```
tensor([[[ 0.00,  0.90,  0.10,  0.08,  0.52,  0.92],
         [ 1.00,  0.90,  0.55,  0.20,  0.90,  0.88],
         [-1.00,  0.80,  0.08,  0.20,  0.56,  0.95],
         [-1.00,  0.70,  0.15,  0.30,  0.62,  0.91]]])
```

After removing those predicted bounding boxes of class -1, we can output the final predicted bounding box kept by non-maximum suppression.

```
fig = d2l.plt.imshow(img)
for i in output[0].detach().numpy():
    if i[0] == -1:
        continue
    label = ('dog=', 'cat=')[int(i[0])] + str(i[1])
    show_bboxes(fig.axes, [torch.tensor(i[2:]) * bbox_scale], label)
```

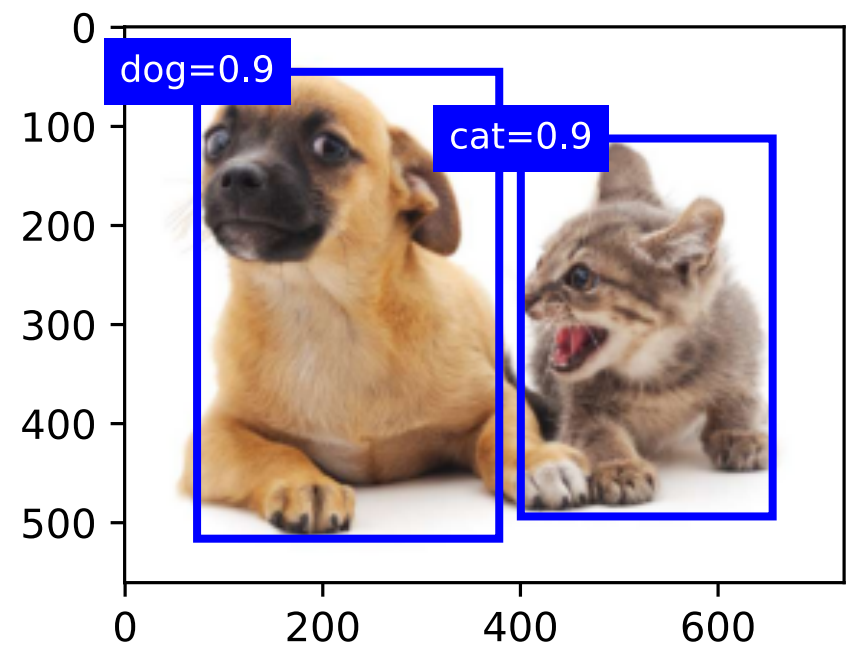

In practice, we can remove predicted bounding boxes with lower confidence even before performing non-maximum suppression, thereby reducing computation in this algorithm. We may also post-process the output of non-maximum suppression, for example, by only keeping results with higher confidence in the final output.

## 14.4.5 Summary

- We generate anchor boxes with different shapes centered on each pixel of the image.

- Intersection over union (IoU), also known as Jaccard index, measures the similarity of two bounding boxes. It is the ratio of their intersection area to their union area.
- In a training set, we need two types of labels for each anchor box. One is the class of the object relevant to the anchor box and the other is the offset of the ground-truth bounding box relative to the anchor box.
- During prediction, we can use non-maximum suppression (NMS) to remove similar predicted bounding boxes, thereby simplifying the output.

### 14.4.6 Exercises

1. Change values of `sizes` and `ratios` in the `multibox_prior` function. What are the changes to the generated anchor boxes?
2. Construct and visualize two bounding boxes with an IoU of 0.5. How do they overlap with each other?
3. Modify the variable `anchors` in Section 14.4.3 and Section 14.4.4. How do the results change?
4. Non-maximum suppression is a greedy algorithm that suppresses predicted bounding boxes by *removing* them. Is it possible that some of these removed ones are actually useful? How can this algorithm be modified to suppress *softly*? You may refer to Soft-NMS (Bodla *et al.*, 2017).
5. Rather than being hand-crafted, can non-maximum suppression be learned?

Discussions[215].

## 14.5 Multiscale Object Detection

In Section 14.4, we generated multiple anchor boxes centered on each pixel of an input image. Essentially these anchor boxes represent samples of different regions of the image. However, we may end up with too many anchor boxes to compute if they are generated for *every* pixel. Think of a $561 \times 728$ input image. If five anchor boxes with varying shapes are generated for each pixel as their center, over two million anchor boxes ($561 \times 728 \times 5$) need to be labeled and predicted on the image.

### 14.5.1 Multiscale Anchor Boxes

You may realize that it is not difficult to reduce anchor boxes on an image. For instance, we can just uniformly sample a small portion of pixels from the input image to generate anchor boxes centered on them. In addition, at different scales we can generate different numbers of anchor boxes of different sizes. Intuitively, smaller objects are more likely to appear on an image than larger ones. As an example, $1 \times 1$, $1 \times 2$, and $2 \times 2$ objects can appear on a $2 \times 2$ image in 4, 2, and 1 possible ways, respectively. Therefore, when using smaller anchor boxes to detect smaller objects, we can sample more regions, while for larger objects we can sample fewer regions.

To demonstrate how to generate anchor boxes at multiple scales, let's read an image. Its height and width are 561 and 728 pixels, respectively.

```
%matplotlib inline
import torch
from d2l import torch as d2l

img = d2l.plt.imread('../img/catdog.jpg')
h, w = img.shape[:2]
h, w
```

```
(561, 728)
```

Recall that in Section 7.2 we call a two-dimensional array output of a convolutional layer a feature map. By defining the feature map shape, we can determine centers of uniformly sampled anchor boxes on any image.

The `display_anchors` function is defined below. We generate anchor boxes (`anchors`) on the feature map (`fmap`) with each unit (pixel) as the anchor box center. Since the $(x, y)$-axis coordinate values in the anchor boxes (`anchors`) have been divided by the width and height of the feature map (`fmap`), these values are between 0 and 1, which indicate the relative positions of anchor boxes in the feature map.

Since centers of the anchor boxes (`anchors`) are spread over all units on the feature map (`fmap`), these centers must be *uniformly* distributed on any input image in terms of their relative spatial positions. More concretely, given the width and height of the feature map `fmap_w` and `fmap_h`, respectively, the following function will *uniformly* sample pixels in `fmap_h` rows and `fmap_w` columns on any input image. Centered on these uniformly sampled pixels, anchor boxes of scale `s` (assuming the length of the list `s` is 1) and different aspect ratios (`ratios`) will be generated.

```
def display_anchors(fmap_w, fmap_h, s):
    d2l.set_figsize()
    # Values on the first two dimensions do not affect the output
    fmap = torch.zeros((1, 10, fmap_h, fmap_w))
    anchors = d2l.multibox_prior(fmap, sizes=s, ratios=[1, 2, 0.5])
    bbox_scale = torch.tensor((w, h, w, h))
    d2l.show_bboxes(d2l.plt.imshow(img).axes,
                    anchors[0] * bbox_scale)
```

First, let's consider detection of small objects. In order to make it easier to distinguish when displayed, the anchor boxes with different centers here do not overlap: the anchor box scale is set to 0.15 and the height and width of the feature map are set to 4. We can see that the centers of the anchor boxes in 4 rows and 4 columns on the image are uniformly distributed.

```
display_anchors(fmap_w=4, fmap_h=4, s=[0.15])
```

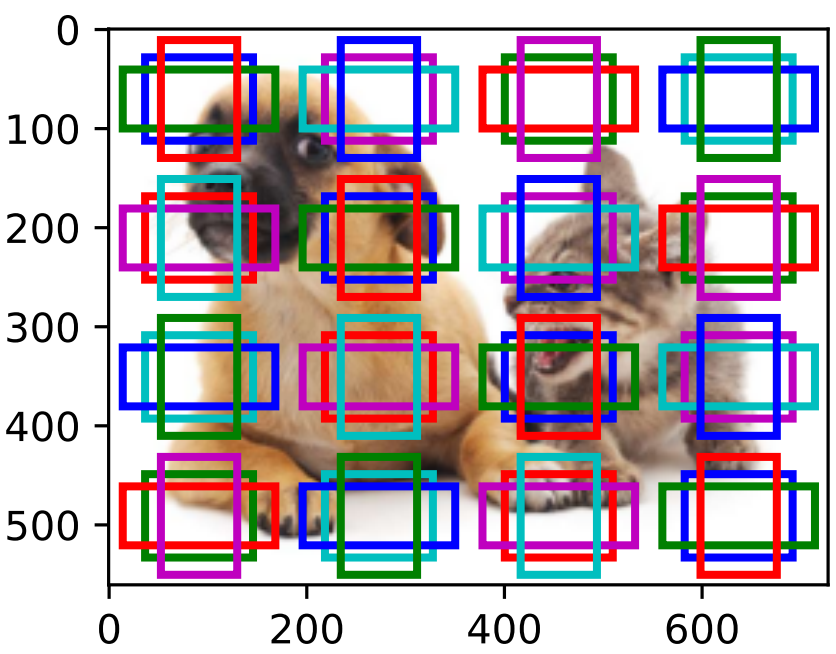

We move on to reduce the height and width of the feature map by half and use larger anchor boxes to detect larger objects. When the scale is set to 0.4, some anchor boxes will overlap with each other.

```
display_anchors(fmap_w=2, fmap_h=2, s=[0.4])
```

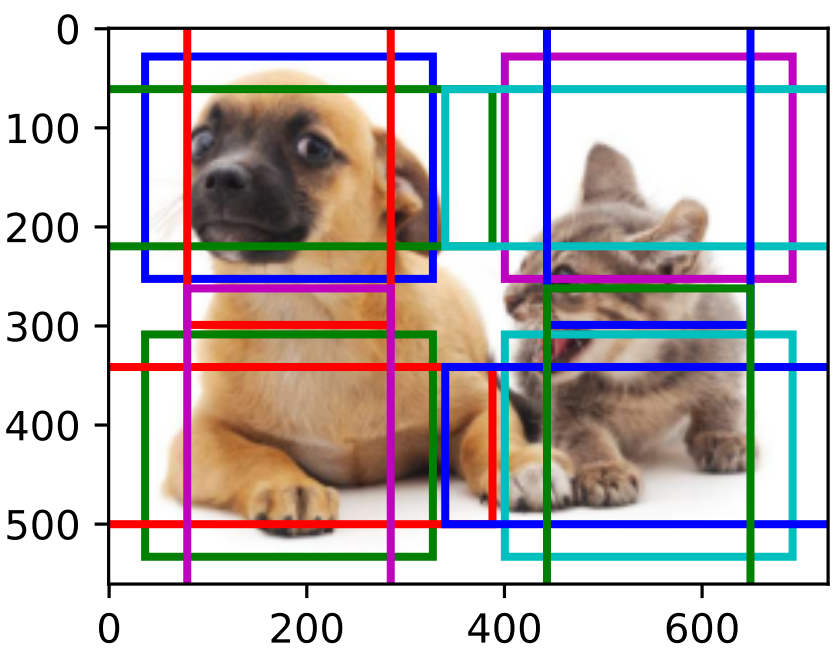

Finally, we further reduce the height and width of the feature map by half and increase the anchor box scale to 0.8. Now the center of the anchor box is the center of the image.

```
display_anchors(fmap_w=1, fmap_h=1, s=[0.8])
```

### 14.5.2 Multiscale Detection

Since we have generated multiscale anchor boxes, we will use them to detect objects of various sizes at different scales. In the following we introduce a CNN-based multiscale object detection method that we will implement in Section 14.7.

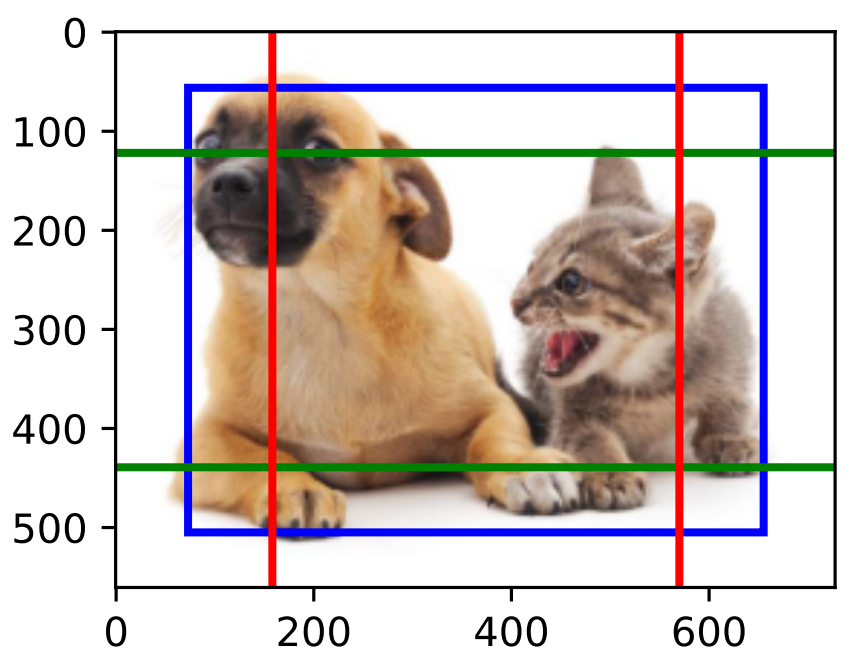

At some scale, say that we have $c$ feature maps of shape $h \times w$. Using the method in Section 14.5.1, we generate $hw$ sets of anchor boxes, where each set has $a$ anchor boxes with the same center. For example, at the first scale in the experiments in Section 14.5.1, given ten (number of channels) $4 \times 4$ feature maps, we generated 16 sets of anchor boxes, where each set contains 3 anchor boxes with the same center. Next, each anchor box is labeled with the class and offset based on ground-truth bounding boxes. At the current scale, the object detection model needs to predict the classes and offsets of $hw$ sets of anchor boxes on the input image, where different sets have different centers.

Assume that the $c$ feature maps here are the intermediate outputs obtained by the CNN forward propagation based on the input image. Since there are $hw$ different spatial positions on each feature map, the same spatial position can be thought of as having $c$ units. According to the definition of receptive field in Section 7.2, these $c$ units at the same spatial position of the feature maps have the same receptive field on the input image: they represent the input image information in the same receptive field. Therefore, we can transform the $c$ units of the feature maps at the same spatial position into the classes and offsets of the $a$ anchor boxes generated using this spatial position. In essence, we use the information of the input image in a certain receptive field to predict the classes and offsets of the anchor boxes that are close to that receptive field on the input image.

When the feature maps at different layers have varying-size receptive fields on the input image, they can be used to detect objects of different sizes. For example, we can design a neural network where units of feature maps that are closer to the output layer have wider receptive fields, so they can detect larger objects from the input image.

In a nutshell, we can leverage layerwise representations of images at multiple levels by deep neural networks for multiscale object detection. We will show how this works through a concrete example in Section 14.7.

### 14.5.3 Summary

- At multiple scales, we can generate anchor boxes with different sizes to detect objects with different sizes.

- By defining the shape of feature maps, we can determine centers of uniformly sampled anchor boxes on any image.
- We use the information of the input image in a certain receptive field to predict the classes and offsets of the anchor boxes that are close to that receptive field on the input image.
- Through deep learning, we can leverage its layerwise representations of images at multiple levels for multiscale object detection.

### 14.5.4 Exercises

1. According to our discussions in Section 8.1, deep neural networks learn hierarchical features with increasing levels of abstraction for images. In multiscale object detection, do feature maps at different scales correspond to different levels of abstraction? Why or why not?
2. At the first scale (`fmap_w=4, fmap_h=4`) in the experiments in Section 14.5.1, generate uniformly distributed anchor boxes that may overlap.
3. Given a feature map variable with shape $1 \times c \times h \times w$, where $c$, $h$, and $w$ are the number of channels, height, and width of the feature maps, respectively. How can you transform this variable into the classes and offsets of anchor boxes? What is the shape of the output?

Discussions[216].

## 14.6 The Object Detection Dataset

There is no small dataset such as MNIST and Fashion-MNIST in the field of object detection. In order to quickly demonstrate object detection models, we collected and labeled a small dataset. First, we took photos of free bananas from our office and generated 1000 banana images with different rotations and sizes. Then we placed each banana image at a random position on some background image. In the end, we labeled bounding boxes for those bananas on the images.

### 14.6.1 Downloading the Dataset

The banana detection dataset with all the image and csv label files can be downloaded directly from the Internet.

```
%matplotlib inline
import os
import pandas as pd
import torch
import torchvision
from d2l import torch as d2l
```

```
#@save
d2l.DATA_HUB['banana-detection'] = (
    d2l.DATA_URL + 'banana-detection.zip',
    '5de26c8fce5ccdea9f91267273464dc968d20d72')
```

### 14.6.2 Reading the Dataset

We are going to read the banana detection dataset in the `read_data_bananas` function below. The dataset includes a csv file for object class labels and ground-truth bounding box coordinates at the upper-left and lower-right corners.

```
#@save
def read_data_bananas(is_train=True):
    """Read the banana detection dataset images and labels."""
    data_dir = d2l.download_extract('banana-detection')
    csv_fname = os.path.join(data_dir, 'bananas_train' if is_train
                             else 'bananas_val', 'label.csv')
    csv_data = pd.read_csv(csv_fname)
    csv_data = csv_data.set_index('img_name')
    images, targets = [], []
    for img_name, target in csv_data.iterrows():
        images.append(torchvision.io.read_image(
            os.path.join(data_dir, 'bananas_train' if is_train else
                         'bananas_val', 'images', f'{img_name}')))
        # Here `target` contains (class, upper-left x, upper-left y,
        # lower-right x, lower-right y), where all the images have the same
        # banana class (index 0)
        targets.append(list(target))
    return images, torch.tensor(targets).unsqueeze(1) / 256
```

By using the `read_data_bananas` function to read images and labels, the following `BananasDataset` class will allow us to create a customized `Dataset` instance for loading the banana detection dataset.

```
#@save
class BananasDataset(torch.utils.data.Dataset):
    """A customized dataset to load the banana detection dataset."""
    def __init__(self, is_train):
        self.features, self.labels = read_data_bananas(is_train)
        print('read ' + str(len(self.features)) + (f' training examples' if
              is_train else f' validation examples'))
```

```
    def __getitem__(self, idx):
        return (self.features[idx].float(), self.labels[idx])

    def __len__(self):
        return len(self.features)
```

Finally, we define the `load_data_bananas` function to return two data iterator instances for both the training and test sets. For the test dataset, there is no need to read it in random order.

```
#@save
def load_data_bananas(batch_size):
    """Load the banana detection dataset."""
    train_iter = torch.utils.data.DataLoader(BananasDataset(is_train=True),
                                             batch_size, shuffle=True)
    val_iter = torch.utils.data.DataLoader(BananasDataset(is_train=False),
                                           batch_size)
    return train_iter, val_iter
```

Let's read a minibatch and print the shapes of both images and labels in this minibatch. The shape of the image minibatch, (batch size, number of channels, height, width), looks familiar: it is the same as in our earlier image classification tasks. The shape of the label minibatch is (batch size, $m$, 5), where $m$ is the largest possible number of bounding boxes that any image has in the dataset.

Although computation in minibatches is more efficient, it requires that all the image examples contain the same number of bounding boxes to form a minibatch via concatenation. In general, images may have a varying number of bounding boxes; thus, images with fewer than $m$ bounding boxes will be padded with illegal bounding boxes until $m$ is reached. Then the label of each bounding box is represented by an array of length 5. The first element in the array is the class of the object in the bounding box, where -1 indicates an illegal bounding box for padding. The remaining four elements of the array are the ($x$, $y$)-coordinate values of the upper-left corner and the lower-right corner of the bounding box (the range is between 0 and 1). For the banana dataset, since there is only one bounding box on each image, we have $m = 1$.

```
batch_size, edge_size = 32, 256
train_iter, _ = load_data_bananas(batch_size)
batch = next(iter(train_iter))
batch[0].shape, batch[1].shape
```

```
Downloading ../data/banana-detection.zip from http://d2l-data.s3-accelerate.
↪amazonaws.com/banana-detection.zip...
read 1000 training examples
read 100 validation examples
```

```
(torch.Size([32, 3, 256, 256]), torch.Size([32, 1, 5]))
```

### 14.6.3 Demonstration

Let's demonstrate ten images with their labeled ground-truth bounding boxes. We can see that the rotations, sizes, and positions of bananas vary across all these images. Of course, this is just a simple artificial dataset. In practice, real-world datasets are usually much more complicated.

```
imgs = (batch[0][:10].permute(0, 2, 3, 1)) / 255
axes = d2l.show_images(imgs, 2, 5, scale=2)
for ax, label in zip(axes, batch[1][:10]):
    d2l.show_bboxes(ax, [label[0][1:5] * edge_size], colors=['w'])
```

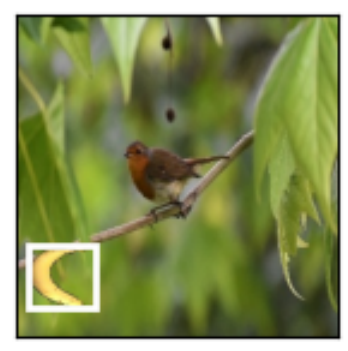
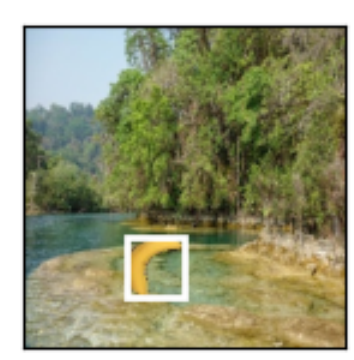
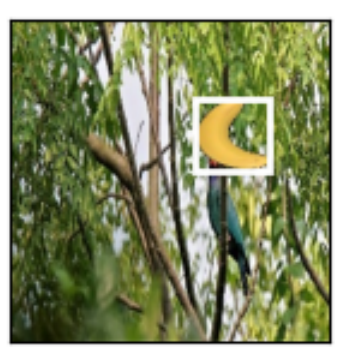
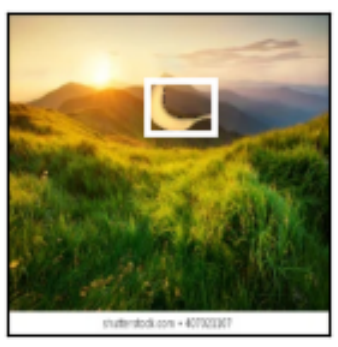
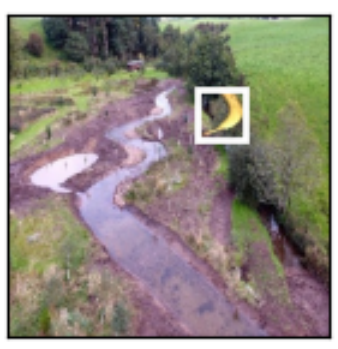
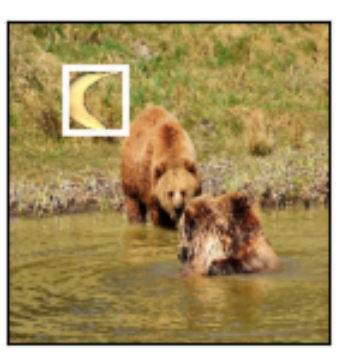
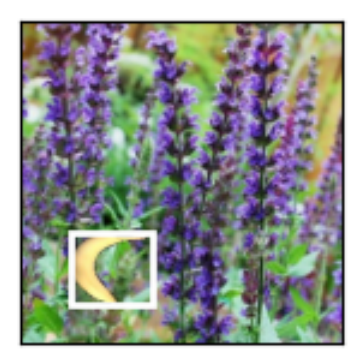
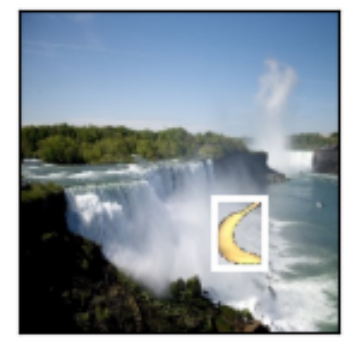
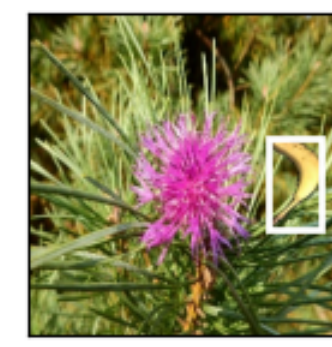
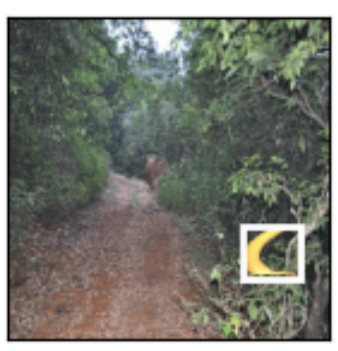

### 14.6.4 Summary

- The banana detection dataset we collected can be used to demonstrate object detection models.
- The data loading for object detection is similar to that for image classification. However, in object detection the labels also contain information of ground-truth bounding boxes, which is missing in image classification.

### 14.6.5 Exercises

1. Demonstrate other images with ground-truth bounding boxes in the banana detection dataset. How do they differ with respect to bounding boxes and objects?
2. Say that we want to apply data augmentation, such as random cropping, to object detection. How can it be different from that in image classification? Hint: what if a cropped image only contains a small portion of an object?

Discussions[217].

## 14.7 Single Shot Multibox Detection

In Section 14.3–Section 14.6, we introduced bounding boxes, anchor boxes, multiscale object detection, and the dataset for object detection. Now we are ready to use such background knowledge to design an object detection model: single shot multibox detection (SSD) (Liu *et al.*, 2016). This model is simple, fast, and widely used. Although this is just one of vast amounts of object detection models, some of the design principles and implementation details in this section are also applicable to other models.

### 14.7.1 Model

Fig. 14.7.1 provides an overview of the design of single-shot multibox detection. This model mainly consists of a base network followed by several multiscale feature map blocks. The base network is for extracting features from the input image, so it can use a deep CNN. For example, the original single-shot multibox detection paper adopts a VGG network truncated before the classification layer (Liu *et al.*, 2016), while ResNet has also been commonly used. Through our design we can make the base network output larger feature maps so as to generate more anchor boxes for detecting smaller objects. Subsequently, each multiscale feature map block reduces (e.g., by half) the height and width of the feature maps from the previous block, and enables each unit of the feature maps to increase its receptive field on the input image.

Recall the design of multiscale object detection through layerwise representations of images by deep neural networks in Section 14.5. Since multiscale feature maps closer to the top of Fig. 14.7.1 are smaller but have larger receptive fields, they are suitable for detecting fewer but larger objects.

In a nutshell, via its base network and several multiscale feature map blocks, single-shot multibox detection generates a varying number of anchor boxes with different sizes, and detects varying-size objects by predicting classes and offsets of these anchor boxes (thus the bounding boxes); thus, this is a multiscale object detection model.

In the following, we will describe the implementation details of different blocks in Fig. 14.7.1. To begin with, we discuss how to implement the class and bounding box prediction.

#### Class Prediction Layer

Let the number of object classes be $q$. Then anchor boxes have $q + 1$ classes, where class 0 is background. At some scale, suppose that the height and width of feature maps are $h$ and $w$, respectively. When $a$ anchor boxes are generated with each spatial position of these feature maps as their center, a total of $hwa$ anchor boxes need to be classified. This often makes classification with fully connected layers infeasible due to likely heavy parametrization costs. Recall how we used channels of convolutional layers to predict classes in Section 8.3. Single-shot multibox detection uses the same technique to reduce model complexity.

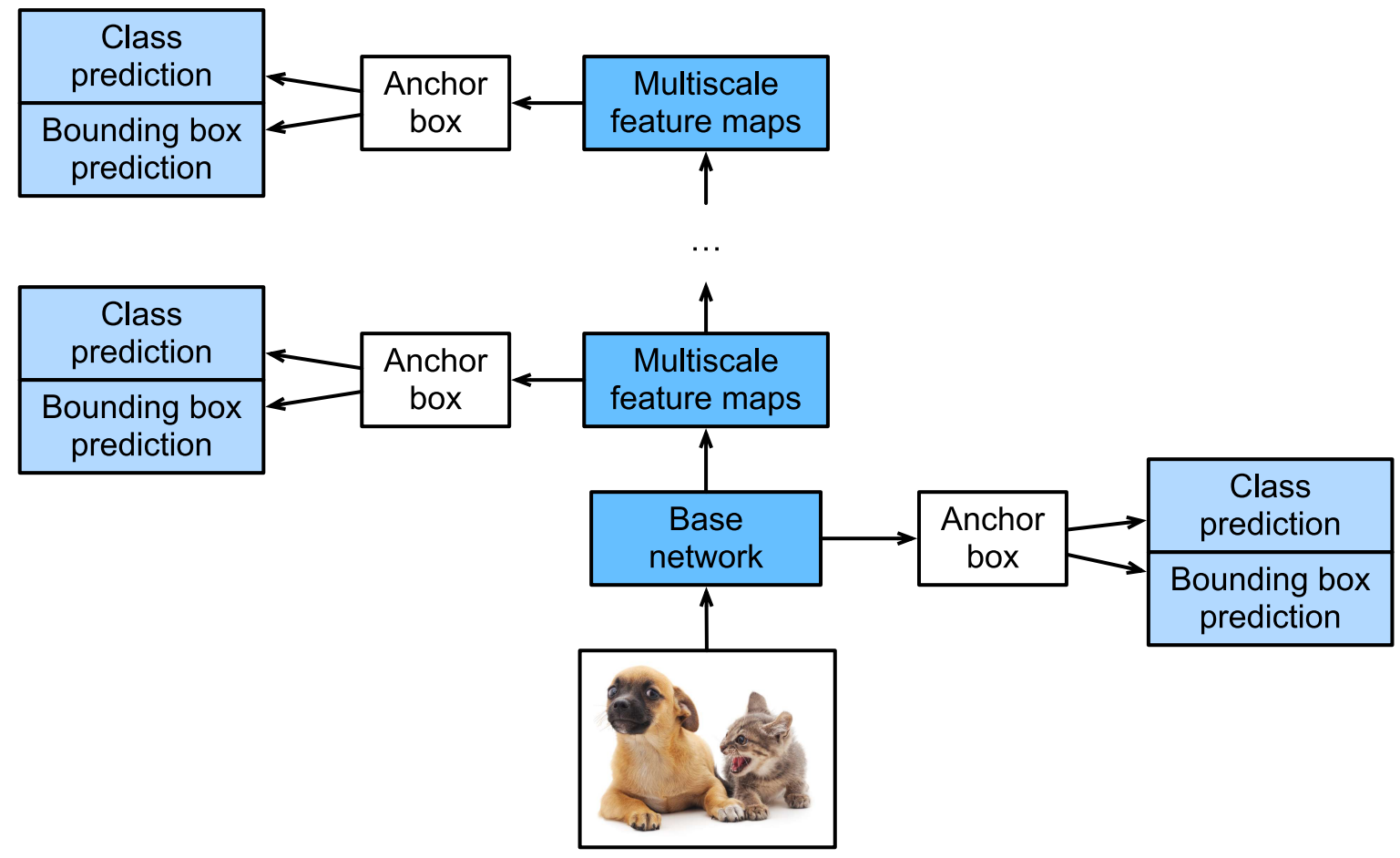

Fig. 14.7.1 As a multiscale object detection model, single-shot multibox detection mainly consists of a base network followed by several multiscale feature map blocks.

Specifically, the class prediction layer uses a convolutional layer without altering width or height of feature maps. In this way, there can be a one-to-one correspondence between outputs and inputs at the same spatial dimensions (width and height) of feature maps. More concretely, channels of the output feature maps at any spatial position $(x, y)$ represent class predictions for all the anchor boxes centered on $(x, y)$ of the input feature maps. To produce valid predictions, there must be $a(q+1)$ output channels, where for the same spatial position the output channel with index $i(q+1)+j$ represents the prediction of the class $j$ $(0 \leq j \leq q)$ for the anchor box $i$ $(0 \leq i < a)$.

Below we define such a class prediction layer, specifying $a$ and $q$ via arguments `num_anchors` and `num_classes`, respectively. This layer uses a $3 \times 3$ convolutional layer with a padding of 1. The width and height of the input and output of this convolutional layer remain unchanged.

```
%matplotlib inline
import torch
import torchvision
from torch import nn
from torch.nn import functional as F
from d2l import torch as d2l

def cls_predictor(num_inputs, num_anchors, num_classes):
```

```
    return nn.Conv2d(num_inputs, num_anchors * (num_classes + 1),
                     kernel_size=3, padding=1)
```

### Bounding Box Prediction Layer

The design of the bounding box prediction layer is similar to that of the class prediction layer. The only difference lies in the number of outputs for each anchor box: here we need to predict four offsets rather than $q + 1$ classes.

```
def bbox_predictor(num_inputs, num_anchors):
    return nn.Conv2d(num_inputs, num_anchors * 4, kernel_size=3, padding=1)
```

### Concatenating Predictions for Multiple Scales

As we mentioned, single-shot multibox detection uses multiscale feature maps to generate anchor boxes and predict their classes and offsets. At different scales, the shapes of feature maps or the numbers of anchor boxes centered on the same unit may vary. Therefore, shapes of the prediction outputs at different scales may vary.

In the following example, we construct feature maps at two different scales, `Y1` and `Y2`, for the same minibatch, where the height and width of `Y2` are half of those of `Y1`. Let's take class prediction as an example. Suppose that 5 and 3 anchor boxes are generated for every unit in `Y1` and `Y2`, respectively. Suppose further that the number of object classes is 10. For feature maps `Y1` and `Y2` the numbers of channels in the class prediction outputs are $5 \times (10 + 1) = 55$ and $3 \times (10 + 1) = 33$, respectively, where either output shape is (batch size, number of channels, height, width).

```
def forward(x, block):
    return block(x)

Y1 = forward(torch.zeros((2, 8, 20, 20)), cls_predictor(8, 5, 10))
Y2 = forward(torch.zeros((2, 16, 10, 10)), cls_predictor(16, 3, 10))
Y1.shape, Y2.shape
```

```
(torch.Size([2, 55, 20, 20]), torch.Size([2, 33, 10, 10]))
```

As we can see, except for the batch size dimension, the other three dimensions all have different sizes. To concatenate these two prediction outputs for more efficient computation, we will transform these tensors into a more consistent format.

Note that the channel dimension holds the predictions for anchor boxes with the same center. We first move this dimension to the innermost. Since the batch size remains the same for different scales, we can transform the prediction output into a two-dimensional tensor with shape (batch size, height × width × number of channels). Then we can concatenate such outputs at different scales along dimension 1.

```
def flatten_pred(pred):
    return torch.flatten(pred.permute(0, 2, 3, 1), start_dim=1)

def concat_preds(preds):
    return torch.cat([flatten_pred(p) for p in preds], dim=1)
```

In this way, even though `Y1` and `Y2` have different sizes in channels, heights, and widths, we can still concatenate these two prediction outputs at two different scales for the same minibatch.

```
concat_preds([Y1, Y2]).shape
```

```
torch.Size([2, 25300])
```

### Downsampling Block

In order to detect objects at multiple scales, we define the following downsampling block `down_sample_blk` that halves the height and width of input feature maps. In fact, this block applies the design of VGG blocks in Section 8.2.1. More concretely, each downsampling block consists of two $3 \times 3$ convolutional layers with padding of 1 followed by a $2 \times 2$ max-pooling layer with stride of 2. As we know, $3 \times 3$ convolutional layers with padding of 1 do not change the shape of feature maps. However, the subsequent $2 \times 2$ max-pooling reduces the height and width of input feature maps by half. For both input and output feature maps of this downsampling block, because $1 \times 2 + (3-1) + (3-1) = 6$, each unit in the output has a $6 \times 6$ receptive field on the input. Therefore, the downsampling block enlarges the receptive field of each unit in its output feature maps.

```
def down_sample_blk(in_channels, out_channels):
    blk = []
    for _ in range(2):
        blk.append(nn.Conv2d(in_channels, out_channels,
                             kernel_size=3, padding=1))
        blk.append(nn.BatchNorm2d(out_channels))
        blk.append(nn.ReLU())
        in_channels = out_channels
    blk.append(nn.MaxPool2d(2))
    return nn.Sequential(*blk)
```

In the following example, our constructed downsampling block changes the number of input channels and halves the height and width of the input feature maps.

```
forward(torch.zeros((2, 3, 20, 20)), down_sample_blk(3, 10)).shape
```

```
torch.Size([2, 10, 10, 10])
```

### Base Network Block

The base network block is used to extract features from input images. For simplicity, we construct a small base network consisting of three downsampling blocks that double the number of channels at each block. Given a $256 \times 256$ input image, this base network block outputs $32 \times 32$ feature maps ($256/2^3 = 32$).

```
def base_net():
    blk = []
    num_filters = [3, 16, 32, 64]
    for i in range(len(num_filters) - 1):
        blk.append(down_sample_blk(num_filters[i], num_filters[i+1]))
    return nn.Sequential(*blk)

forward(torch.zeros((2, 3, 256, 256)), base_net()).shape
```

```
torch.Size([2, 64, 32, 32])
```

### The Complete Model

The complete single shot multibox detection model consists of five blocks. The feature maps produced by each block are used for both (i) generating anchor boxes and (ii) predicting classes and offsets of these anchor boxes. Among these five blocks, the first one is the base network block, the second to the fourth are downsampling blocks, and the last block uses global max-pooling to reduce both the height and width to 1. Technically, the second to the fifth blocks are all those multiscale feature map blocks in Fig. 14.7.1.

```
def get_blk(i):
    if i == 0:
        blk = base_net()
    elif i == 1:
        blk = down_sample_blk(64, 128)
    elif i == 4:
        blk = nn.AdaptiveMaxPool2d((1,1))
    else:
        blk = down_sample_blk(128, 128)
    return blk
```

Now we define the forward propagation for each block. Different from in image classification tasks, outputs here include (i) CNN feature maps `Y`, (ii) anchor boxes generated using `Y` at the current scale, and (iii) classes and offsets predicted (based on `Y`) for these anchor boxes.

```
def blk_forward(X, blk, size, ratio, cls_predictor, bbox_predictor):
    Y = blk(X)
    anchors = d2l.multibox_prior(Y, sizes=size, ratios=ratio)
    cls_preds = cls_predictor(Y)
    bbox_preds = bbox_predictor(Y)
    return (Y, anchors, cls_preds, bbox_preds)
```

Recall that in Fig. 14.7.1 a multiscale feature map block that is closer to the top is for detecting larger objects; thus, it needs to generate larger anchor boxes. In the above forward propagation, at each multiscale feature map block we pass in a list of two scale values via the `sizes` argument of the invoked `multibox_prior` function (described in Section 14.4). In the following, the interval between 0.2 and 1.05 is split evenly into five sections to determine the smaller scale values at the five blocks: 0.2, 0.37, 0.54, 0.71, and 0.88. Then their larger scale values are given by $\sqrt{0.2 \times 0.37} = 0.272$, $\sqrt{0.37 \times 0.54} = 0.447$, and so on.

```
sizes = [[0.2, 0.272], [0.37, 0.447], [0.54, 0.619], [0.71, 0.79],
         [0.88, 0.961]]
ratios = [[1, 2, 0.5]] * 5
num_anchors = len(sizes[0]) + len(ratios[0]) - 1
```

Now we can define the complete model `TinySSD` as follows.

```
class TinySSD(nn.Module):
    def __init__(self, num_classes, **kwargs):
        super(TinySSD, self).__init__(**kwargs)
        self.num_classes = num_classes
        idx_to_in_channels = [64, 128, 128, 128, 128]
        for i in range(5):
            # Equivalent to the assignment statement `self.blk_i = get_blk(i)`
            setattr(self, f'blk_{i}', get_blk(i))
            setattr(self, f'cls_{i}', cls_predictor(idx_to_in_channels[i],
                                                    num_anchors, num_classes))
            setattr(self, f'bbox_{i}', bbox_predictor(idx_to_in_channels[i],
                                                      num_anchors))

    def forward(self, X):
        anchors, cls_preds, bbox_preds = [None] * 5, [None] * 5, [None] * 5
        for i in range(5):
            # Here `getattr(self, 'blk_%d' % i)` accesses `self.blk_i`
            X, anchors[i], cls_preds[i], bbox_preds[i] = blk_forward(
                X, getattr(self, f'blk_{i}'), sizes[i], ratios[i],
                getattr(self, f'cls_{i}'), getattr(self, f'bbox_{i}'))
        anchors = torch.cat(anchors, dim=1)
        cls_preds = concat_preds(cls_preds)
        cls_preds = cls_preds.reshape(
            cls_preds.shape[0], -1, self.num_classes + 1)
        bbox_preds = concat_preds(bbox_preds)
        return anchors, cls_preds, bbox_preds
```

We create a model instance and use it to perform forward propagation on a minibatch of $256 \times 256$ images `X`.

As shown earlier in this section, the first block outputs $32 \times 32$ feature maps. Recall that the second to fourth downsampling blocks halve the height and width and the fifth block uses global pooling. Since 4 anchor boxes are generated for each unit along spatial dimensions of feature maps, at all the five scales a total of $(32^2 + 16^2 + 8^2 + 4^2 + 1) \times 4 = 5444$ anchor boxes are generated for each image.

```
net = TinySSD(num_classes=1)
X = torch.zeros((32, 3, 256, 256))
anchors, cls_preds, bbox_preds = net(X)

print('output anchors:', anchors.shape)
print('output class preds:', cls_preds.shape)
print('output bbox preds:', bbox_preds.shape)
```

```
output anchors: torch.Size([1, 5444, 4])
output class preds: torch.Size([32, 5444, 2])
output bbox preds: torch.Size([32, 21776])
```

## 14.7.2 Training

Now we will explain how to train the single shot multibox detection model for object detection.

### Reading the Dataset and Initializing the Model

To begin with, let's read the banana detection dataset described in Section 14.6.

```
batch_size = 32
train_iter, _ = d2l.load_data_bananas(batch_size)
```

```
read 1000 training examples
read 100 validation examples
```

There is only one class in the banana detection dataset. After defining the model, we need to initialize its parameters and define the optimization algorithm.

```
device, net = d2l.try_gpu(), TinySSD(num_classes=1)
trainer = torch.optim.SGD(net.parameters(), lr=0.2, weight_decay=5e-4)
```

### Defining Loss and Evaluation Functions

Object detection has two types of losses. The first loss concerns classes of anchor boxes: its computation can simply reuse the cross-entropy loss function that we used for image classi-fication. The second loss concerns offsets of positive (non-background) anchor boxes: this is a regression problem. For this regression problem, however, here we do not use the squared loss described in Section 3.1.3. Instead, we use the $\ell_1$ norm loss, the absolute value of the difference between the prediction and the ground-truth. The mask variable `bbox_masks` filters out negative anchor boxes and illegal (padded) anchor boxes in the loss calculation. In the end, we sum up the anchor box class loss and the anchor box offset loss to obtain the loss function for the model.

```
cls_loss = nn.CrossEntropyLoss(reduction='none')
bbox_loss = nn.L1Loss(reduction='none')

def calc_loss(cls_preds, cls_labels, bbox_preds, bbox_labels, bbox_masks):
    batch_size, num_classes = cls_preds.shape[0], cls_preds.shape[2]
    cls = cls_loss(cls_preds.reshape(-1, num_classes),
                   cls_labels.reshape(-1)).reshape(batch_size, -1).mean(dim=1)
    bbox = bbox_loss(bbox_preds * bbox_masks,
                     bbox_labels * bbox_masks).mean(dim=1)
    return cls + bbox
```

We can use accuracy to evaluate the classification results. Due to the used $\ell_1$ norm loss for the offsets, we use the *mean absolute error* to evaluate the predicted bounding boxes. These prediction results are obtained from the generated anchor boxes and the predicted offsets for them.

```
def cls_eval(cls_preds, cls_labels):
    # Because the class prediction results are on the final dimension,
    # `argmax` needs to specify this dimension
    return float((cls_preds.argmax(dim=-1).type(
        cls_labels.dtype) == cls_labels).sum())

def bbox_eval(bbox_preds, bbox_labels, bbox_masks):
    return float((torch.abs((bbox_labels - bbox_preds) * bbox_masks)).sum())
```

### Training the Model

When training the model, we need to generate multiscale anchor boxes (`anchors`) and predict their classes (`cls_preds`) and offsets (`bbox_preds`) in the forward propagation. Then we label the classes (`cls_labels`) and offsets (`bbox_labels`) of such generated anchor boxes based on the label information `Y`. Finally, we calculate the loss function using the predicted and labeled values of the classes and offsets. For concise implementations, evaluation of the test dataset is omitted here.

```
num_epochs, timer = 20, d2l.Timer()
animator = d2l.Animator(xlabel='epoch', xlim=[1, num_epochs],
                        legend=['class error', 'bbox mae'])
net = net.to(device)
for epoch in range(num_epochs):
```

(continued from previous page)

```python
    # Sum of training accuracy, no. of examples in sum of training accuracy,
    # Sum of absolute error, no. of examples in sum of absolute error
    metric = d2l.Accumulator(4)
    net.train()
    for features, target in train_iter:
        timer.start()
        trainer.zero_grad()
        X, Y = features.to(device), target.to(device)
        # Generate multiscale anchor boxes and predict their classes and
        # offsets
        anchors, cls_preds, bbox_preds = net(X)
        # Label the classes and offsets of these anchor boxes
        bbox_labels, bbox_masks, cls_labels = d2l.multibox_target(anchors, Y)
        # Calculate the loss function using the predicted and labeled values
        # of the classes and offsets
        l = calc_loss(cls_preds, cls_labels, bbox_preds, bbox_labels,
                      bbox_masks)
        l.mean().backward()
        trainer.step()
        metric.add(cls_eval(cls_preds, cls_labels), cls_labels.numel(),
                   bbox_eval(bbox_preds, bbox_labels, bbox_masks),
                   bbox_labels.numel())
    cls_err, bbox_mae = 1 - metric[0] / metric[1], metric[2] / metric[3]
    animator.add(epoch + 1, (cls_err, bbox_mae))
print(f'class err {cls_err:.2e}, bbox mae {bbox_mae:.2e}')
print(f'{len(train_iter.dataset) / timer.stop():.1f} examples/sec on '
      f'{str(device)}')
```

```
class err 3.31e-03, bbox mae 3.11e-03
3535.8 examples/sec on cuda:0
```

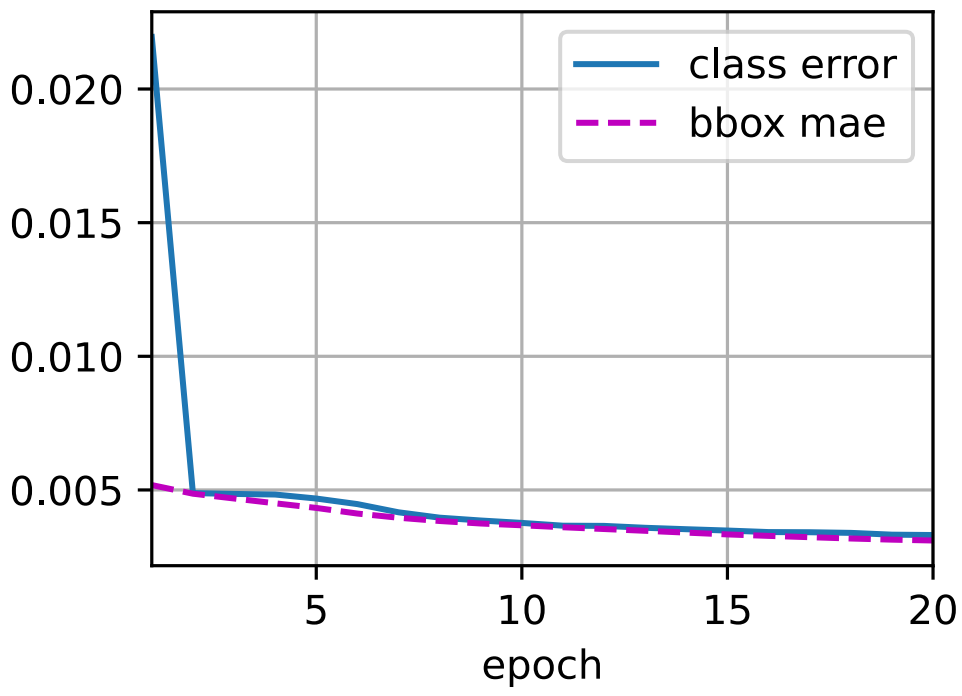

## 14.7.3 Prediction

During prediction, the goal is to detect all the objects of interest on the image. Below we read and resize a test image, converting it to a four-dimensional tensor that is required by convolutional layers.

```
X = torchvision.io.read_image('../img/banana.jpg').unsqueeze(0).float()
img = X.squeeze(0).permute(1, 2, 0).long()
```

Using the `multibox_detection` function below, the predicted bounding boxes are obtained from the anchor boxes and their predicted offsets. Then non-maximum suppression is used to remove similar predicted bounding boxes.

```
def predict(X):
    net.eval()
    anchors, cls_preds, bbox_preds = net(X.to(device))
    cls_probs = F.softmax(cls_preds, dim=2).permute(0, 2, 1)
    output = d2l.multibox_detection(cls_probs, bbox_preds, anchors)
    idx = [i for i, row in enumerate(output[0]) if row[0] != -1]
    return output[0, idx]

output = predict(X)
```

Finally, we display all the predicted bounding boxes with confidence 0.9 or above as output.

```
def display(img, output, threshold):
    d2l.set_figsize((5, 5))
    fig = d2l.plt.imshow(img)
    for row in output:
        score = float(row[1])
        if score < threshold:
            continue
        h, w = img.shape[:2]
        bbox = [row[2:6] * torch.tensor((w, h, w, h), device=row.device)]
        d2l.show_bboxes(fig.axes, bbox, '%.2f' % score, 'w')

display(img, output.cpu(), threshold=0.9)
```

## 14.7.4 Summary

- Single shot multibox detection is a multiscale object detection model. Via its base network and several multiscale feature map blocks, single-shot multibox detection generates a varying number of anchor boxes with different sizes, and detects varying-size objects by predicting classes and offsets of these anchor boxes (thus the bounding boxes).
- When training the single-shot multibox detection model, the loss function is calculated based on the predicted and labeled values of the anchor box classes and offsets.

## 14.7.5 Exercises

1. Can you improve the single-shot multibox detection by improving the loss function? For example, replace $\ell_1$ norm loss with smooth $\ell_1$ norm loss for the predicted offsets. This loss function uses a square function around zero for smoothness, which is controlled by the hyperparameter $\sigma$:

$$f(x) = \begin{cases} (\sigma x)^2/2, & \text{if } |x| < 1/\sigma^2 \\ |x| - 0.5/\sigma^2, & \text{otherwise} \end{cases} \tag{14.7.1}$$

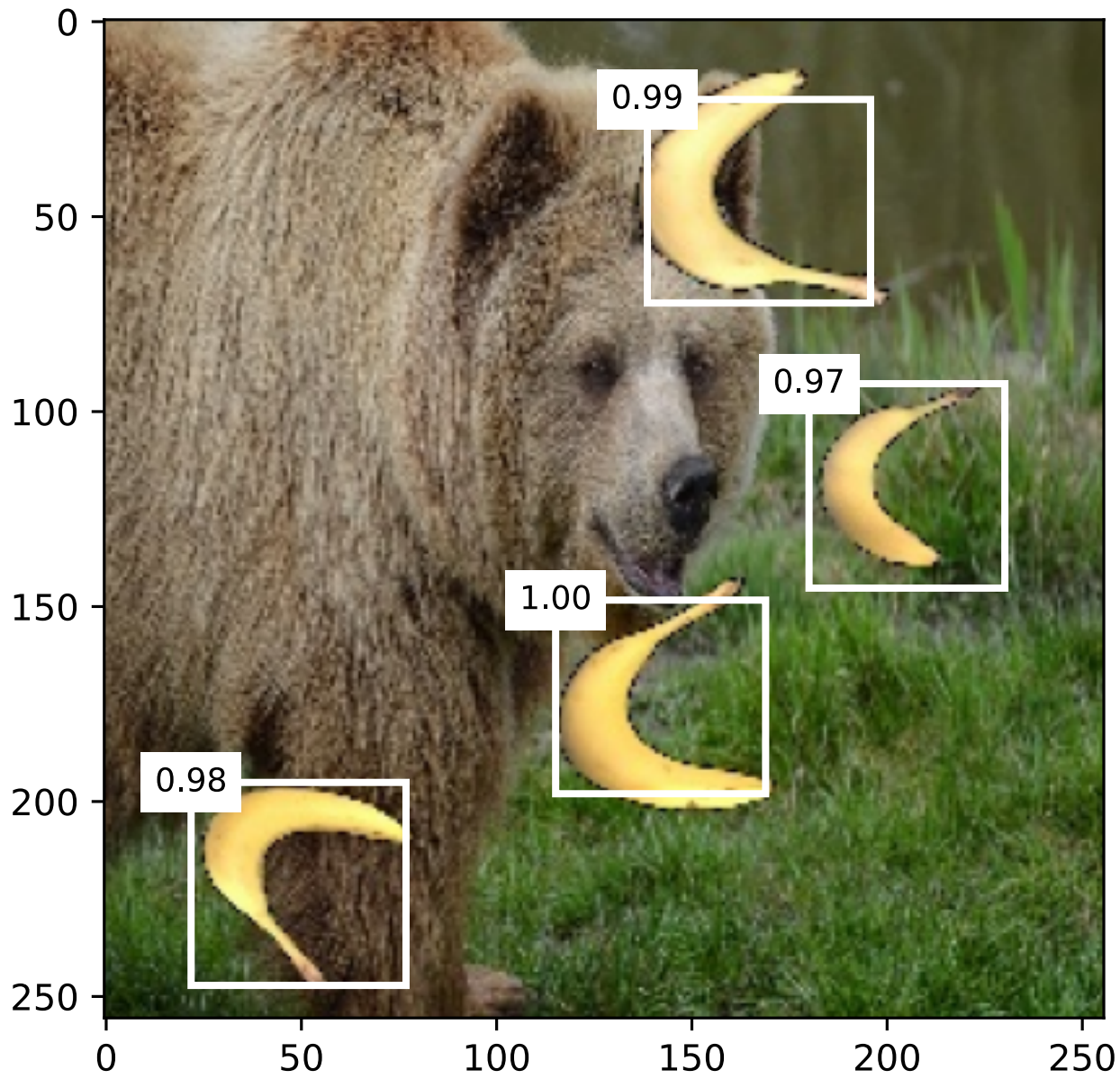

When $\sigma$ is very large, this loss is similar to the $\ell_1$ norm loss. When its value is smaller, the loss function is smoother.

```python
def smooth_l1(data, scalar):
    out = []
    for i in data:
        if abs(i) < 1 / (scalar ** 2):
            out.append(((scalar * i) ** 2) / 2)
        else:
            out.append(abs(i) - 0.5 / (scalar ** 2))
    return torch.tensor(out)

sigmas = [10, 1, 0.5]
lines = ['-', '--', '-.']
x = torch.arange(-2, 2, 0.1)
d2l.set_figsize()

for l, s in zip(lines, sigmas):
    y = smooth_l1(x, scalar=s)
    d2l.plt.plot(x, y, l, label='sigma=%.1f' % s)
d2l.plt.legend();
```

Besides, in the experiment we used cross-entropy loss for class prediction: denoting by $p_j$ the predicted probability for the ground-truth class $j$, the cross-entropy loss is $-\log p_j$. We can also use the focal loss (Lin *et al.*, 2017): given hyperparameters $\gamma > 0$ and $\alpha > 0$, this loss is defined as:

$$-\alpha (1 - p_j)^{\gamma} \log p_j. \tag{14.7.2}$$

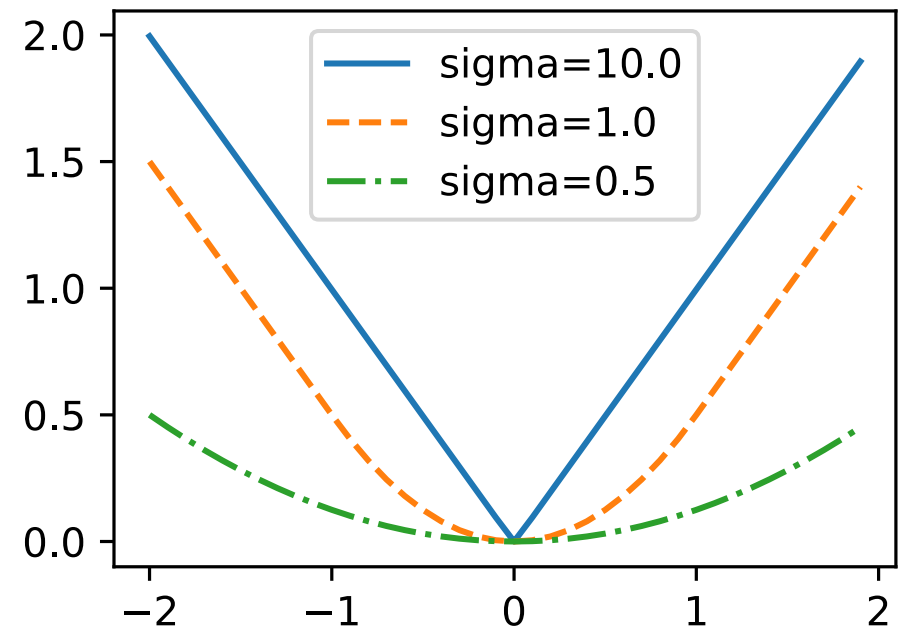

As we can see, increasing $\gamma$ can effectively reduce the relative loss for well-classified examples (e.g., $p_j > 0.5$) so the training can focus more on those difficult examples that are misclassified.

```
def focal_loss(gamma, x):
    return -(1 - x) ** gamma * torch.log(x)

x = torch.arange(0.01, 1, 0.01)
for l, gamma in zip(lines, [0, 1, 5]):
    y = d2l.plt.plot(x, focal_loss(gamma, x), l, label='gamma=%.1f' % gamma)
d2l.plt.legend();
```

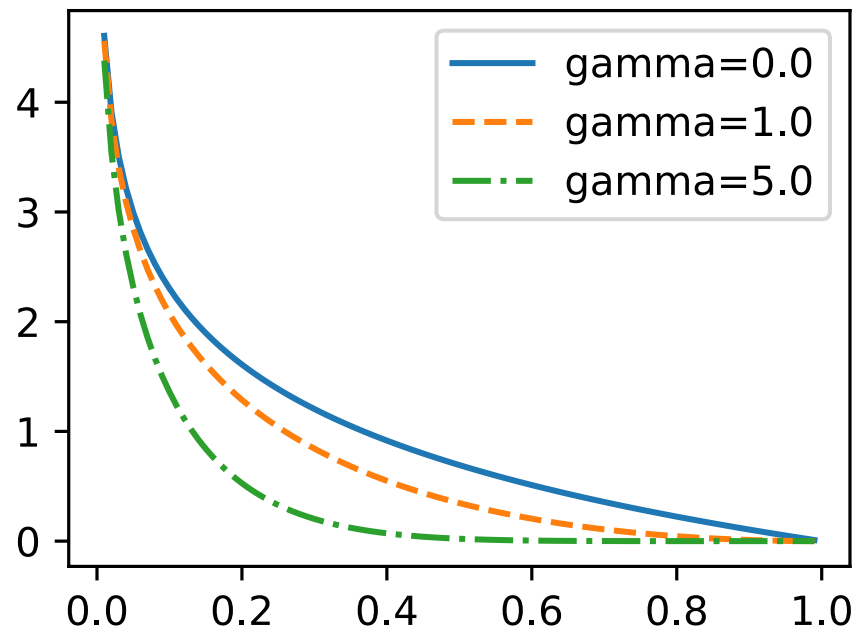

2. Due to space limitations, we have omitted some implementation details of the single shot multibox detection model in this section. Can you further improve the model in the following aspects:

    1. When an object is much smaller compared with the image, the model could resize the input image bigger.

    2. There are typically a vast number of negative anchor boxes. To make the class distribution more balanced, we could downsample negative anchor boxes.

3. In the loss function, assign different weight hyperparameters to the class loss and the offset loss.
4. Use other methods to evaluate the object detection model, such as those in the single shot multibox detection paper (Liu *et al.*, 2016).

Discussions[218].

# 14.8 Region-based CNNs (R-CNNs)

Besides single shot multibox detection described in Section 14.7, region-based CNNs or regions with CNN features (R-CNNs) are also among many pioneering approaches of applying deep learning to object detection (Girshick *et al.*, 2014). In this section, we will introduce the R-CNN and its series of improvements: the fast R-CNN (Girshick, 2015), the faster R-CNN (Ren *et al.*, 2015), and the mask R-CNN (He *et al.*, 2017). Due to limited space, we will only focus on the design of these models.

## 14.8.1 R-CNNs

The *R-CNN* first extracts many (e.g., 2000) *region proposals* from the input image (e.g., anchor boxes can also be considered as region proposals), labeling their classes and bounding boxes (e.g., offsets).

(Girshick *et al.*, 2014)

Then a CNN is used to perform forward propagation on each region proposal to extract its features. Next, features of each region proposal are used for predicting the class and bounding box of this region proposal.

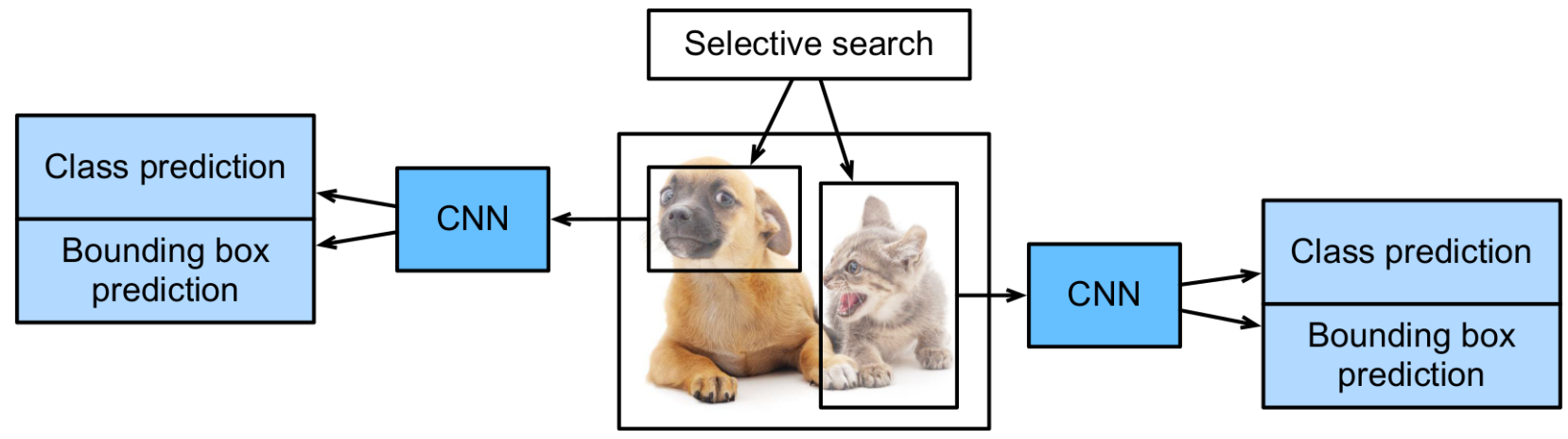

Fig. 14.8.1 The R-CNN model.

Fig. 14.8.1 shows the R-CNN model. More concretely, the R-CNN consists of the following four steps:

1. Perform *selective search* to extract multiple high-quality region proposals on the input image (Uijlings *et al.*, 2013). These proposed regions are usually selected at multiple scales with different shapes and sizes. Each region proposal will be labeled with a class and a ground-truth bounding box.

2. Choose a pretrained CNN and truncate it before the output layer. Resize each region proposal to the input size required by the network, and output the extracted features for the region proposal through forward propagation.

3. Take the extracted features and labeled class of each region proposal as an example. Train multiple support vector machines to classify objects, where each support vector machine individually determines whether the example contains a specific class.

4. Take the extracted features and labeled bounding box of each region proposal as an example. Train a linear regression model to predict the ground-truth bounding box.

Although the R-CNN model uses pretrained CNNs to effectively extract image features, it is slow. Imagine that we select thousands of region proposals from a single input image: this requires thousands of CNN forward propagations to perform object detection. This massive computing load makes it infeasible to widely use R-CNNs in real-world applications.

## 14.8.2 Fast R-CNN

The main performance bottleneck of an R-CNN lies in the independent CNN forward propagation for each region proposal, without sharing computation. Since these regions usually have overlaps, independent feature extractions lead to much repeated computation. One of the major improvements of the *fast R-CNN* from the R-CNN is that the CNN forward propagation is only performed on the entire image (Girshick, 2015).

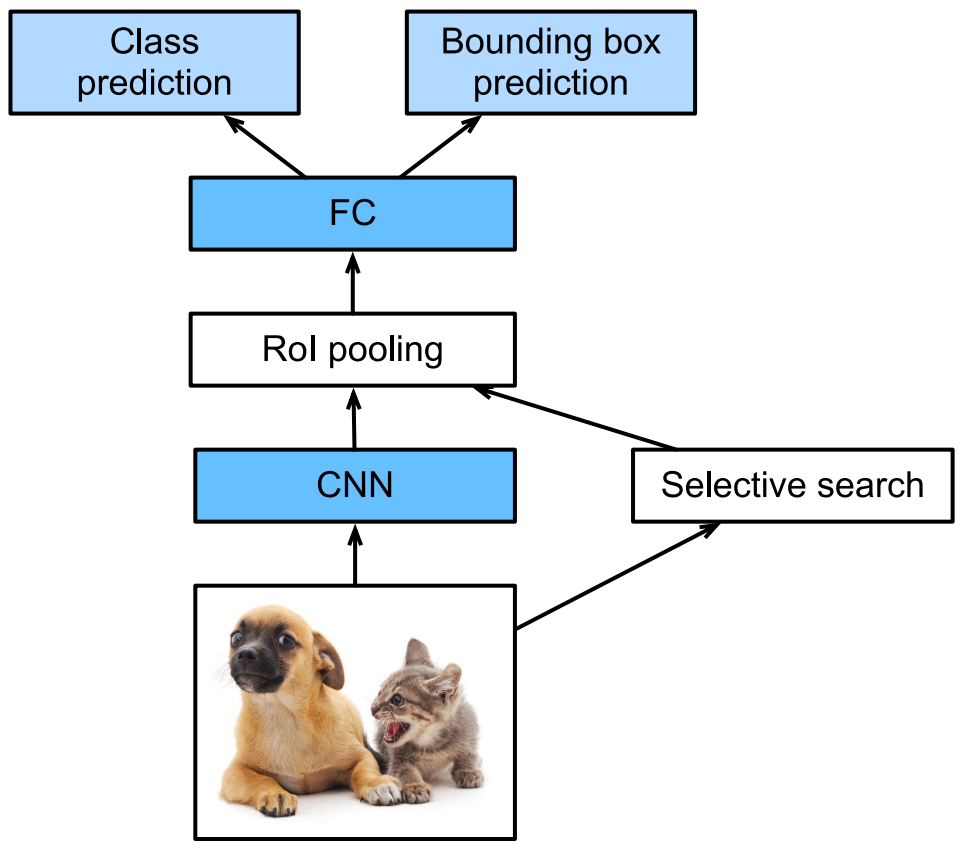

Fig. 14.8.2 The fast R-CNN model.

Fig. 14.8.2 describes the fast R-CNN model. Its major computations are as follows:

1. Compared with the R-CNN, in the fast R-CNN the input of the CNN for feature extraction is the entire image, rather than individual region proposals. Moreover, this CNN is trainable. Given an input image, let the shape of the CNN output be $1 \times c \times h_1 \times w_1$.

2. Suppose that selective search generates $n$ region proposals. These region proposals (of different shapes) mark regions of interest (of different shapes) on the CNN output. Then these regions of interest further extract features of the same shape (say height $h_2$ and width $w_2$ are specified) in order to be easily concatenated. To achieve this, the fast R-CNN introduces the *region of interest (RoI) pooling* layer: the CNN output and region proposals are input into this layer, outputting concatenated features of shape $n \times c \times h_2 \times w_2$ that are further extracted for all the region proposals.

3. Using a fully connected layer, transform the concatenated features into an output of shape $n \times d$, where $d$ depends on the model design.

4. Predict the class and bounding box for each of the $n$ region proposals. More concretely, in class and bounding box prediction, transform the fully connected layer output into an output of shape $n \times q$ ($q$ is the number of classes) and an output of shape $n \times 4$, respectively. The class prediction uses softmax regression.

The region of interest pooling layer proposed in the fast R-CNN is different from the pooling layer introduced in Section 7.5. In the pooling layer, we indirectly control the output shape by specifying sizes of the pooling window, padding, and stride. In contrast, we can directly specify the output shape in the region of interest pooling layer.

For example, let's specify the output height and width for each region as $h_2$ and $w_2$, respectively. For any region of interest window of shape $h \times w$, this window is divided into a $h_2 \times w_2$ grid of subwindows, where the shape of each subwindow is approximately $(h/h_2) \times (w/w_2)$. In practice, the height and width of any subwindow shall be rounded up, and the largest element shall be used as output of the subwindow. Therefore, the region of interest pooling layer can extract features of the same shape even when regions of interest have different shapes.

As an illustrative example, in Fig. 14.8.3, the upper-left $3 \times 3$ region of interest is selected on a $4 \times 4$ input. For this region of interest, we use a $2 \times 2$ region of interest pooling layer to obtain a $2 \times 2$ output. Note that each of the four divided subwindows contains elements 0, 1, 4, and 5 (5 is the maximum); 2 and 6 (6 is the maximum); 8 and 9 (9 is the maximum); and 10.

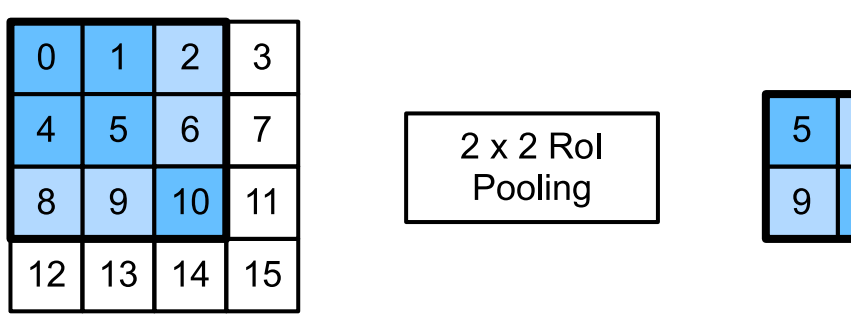

Fig. 14.8.3 A $2 \times 2$ region of interest pooling layer.

Below we demonstrate the computation of the region of interest pooling layer. Suppose that the height and width of the CNN-extracted features `X` are both 4, and there is only a single channel.

```
import torch
import torchvision

X = torch.arange(16.).reshape(1, 1, 4, 4)
X
```

```
tensor([[[[ 0.,  1.,  2.,  3.],
          [ 4.,  5.,  6.,  7.],
          [ 8.,  9., 10., 11.],
          [12., 13., 14., 15.]]]])
```

Let's further suppose that the height and width of the input image are both 40 pixels and that selective search generates two region proposals on this image. Each region proposal is expressed as five elements: its object class followed by the $(x, y)$-coordinates of its upper-left and lower-right corners.

```
rois = torch.Tensor([[0, 0, 0, 20, 20], [0, 0, 10, 30, 30]])
```

Because the height and width of `X` are $1/10$ of the height and width of the input image, the coordinates of the two region proposals are multiplied by 0.1 according to the specified `spatial_scale` argument. Then the two regions of interest are marked on `X` as `X[:, :, 0:3, 0:3]` and `X[:, :, 1:4, 0:4]`, respectively. Finally in the $2 \times 2$ region of interest pooling, each region of interest is divided into a grid of sub-windows to further extract features of the same shape $2 \times 2$.

```
torchvision.ops.roi_pool(X, rois, output_size=(2, 2), spatial_scale=0.1)
```

```
tensor([[[[ 5.,  6.],
          [ 9., 10.]]],

        [[[ 9., 11.],
          [13., 15.]]]])
```

### 14.8.3 Faster R-CNN

To be more accurate in object detection, the fast R-CNN model usually has to generate a lot of region proposals in selective search. To reduce region proposals without loss of accuracy, the *faster R-CNN* proposes to replace selective search with a *region proposal network* (Ren *et al.*, 2015).

Fig. 14.8.4 shows the faster R-CNN model. Compared with the fast R-CNN, the faster R-CNN only changes the region proposal method from selective search to a region proposal network. The rest of the model remain unchanged. The region proposal network works in the following steps:

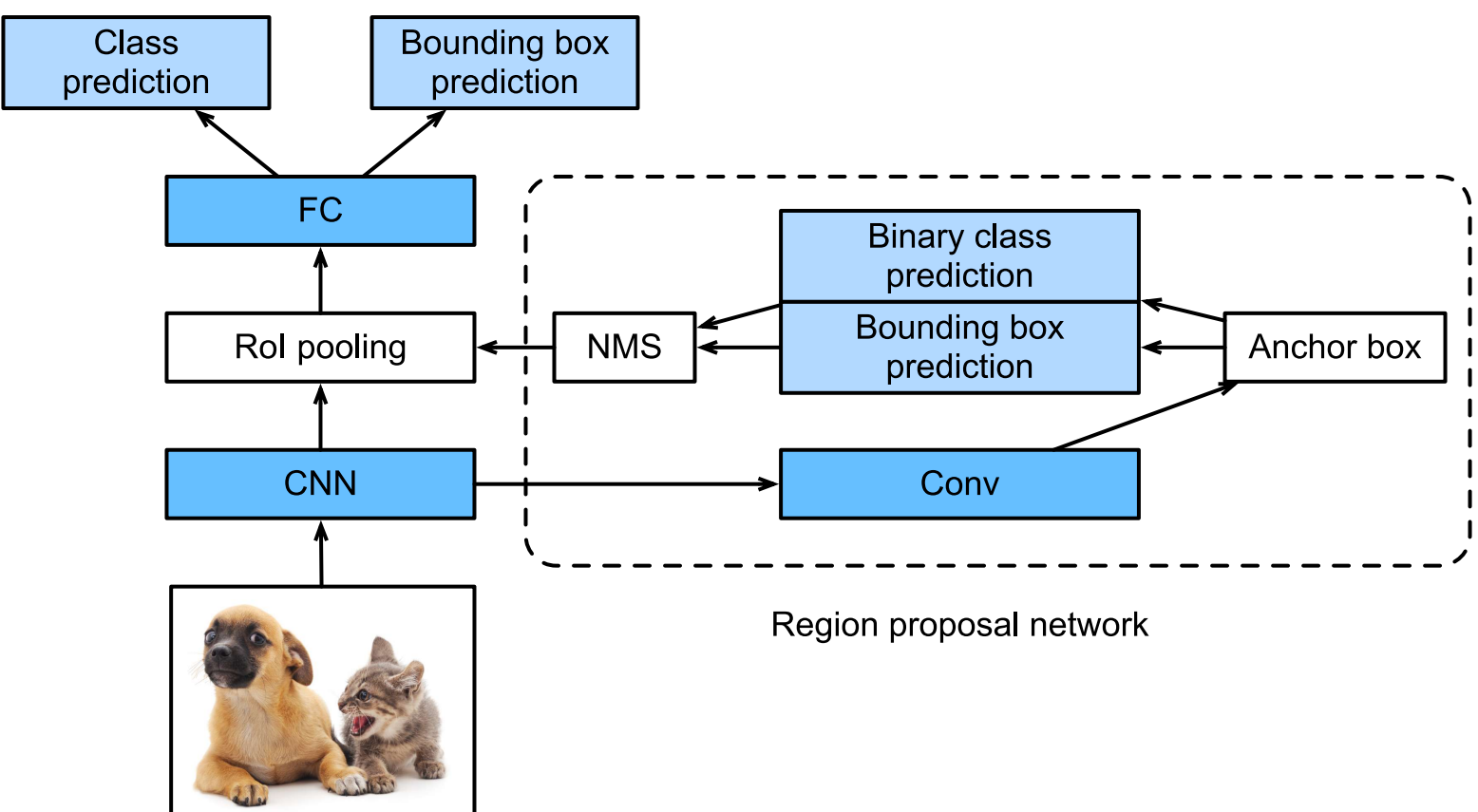

Fig. 14.8.4 The faster R-CNN model.

1. Use a $3 \times 3$ convolutional layer with padding of 1 to transform the CNN output to a new output with $c$ channels. In this way, each unit along the spatial dimensions of the CNN-extracted feature maps gets a new feature vector of length $c$.
2. Centered on each pixel of the feature maps, generate multiple anchor boxes of different scales and aspect ratios and label them.
3. Using the length-$c$ feature vector at the center of each anchor box, predict the binary class (background or objects) and bounding box for this anchor box.
4. Consider those predicted bounding boxes whose predicted classes are objects. Remove overlapped results using non-maximum suppression. The remaining predicted bounding boxes for objects are the region proposals required by the region of interest pooling layer.

It is worth noting that, as part of the faster R-CNN model, the region proposal network is jointly trained with the rest of the model. In other words, the objective function of the faster R-CNN includes not only the class and bounding box prediction in object detection, but also the binary class and bounding box prediction of anchor boxes in the region proposal network. As a result of the end-to-end training, the region proposal network learns how to generate high-quality region proposals, so as to stay accurate in object detection with a reduced number of region proposals that are learned from data.

## 14.8.4 Mask R-CNN

In the training dataset, if pixel-level positions of object are also labeled on images, the *mask R-CNN* can effectively leverage such detailed labels to further improve the accuracy of object detection (He *et al.*, 2017).

As shown in Fig. 14.8.5, the mask R-CNN is modified based on the faster R-CNN. Specifically, the mask R-CNN replaces the region of interest pooling layer with the *region of interest (RoI) alignment* layer. This region of interest alignment layer uses bilinear interpolation to preserve the spatial information on the feature maps, which is more suitable for pixel-level prediction. The output of this layer contains feature maps of the same shape for all the regions of interest. They are used to predict not only the class and bounding box for each region of interest, but also the pixel-level position of the object through an additional fully convolutional network. More details on using a fully convolutional network to predict pixel-level semantics of an image will be provided in subsequent sections of this chapter.

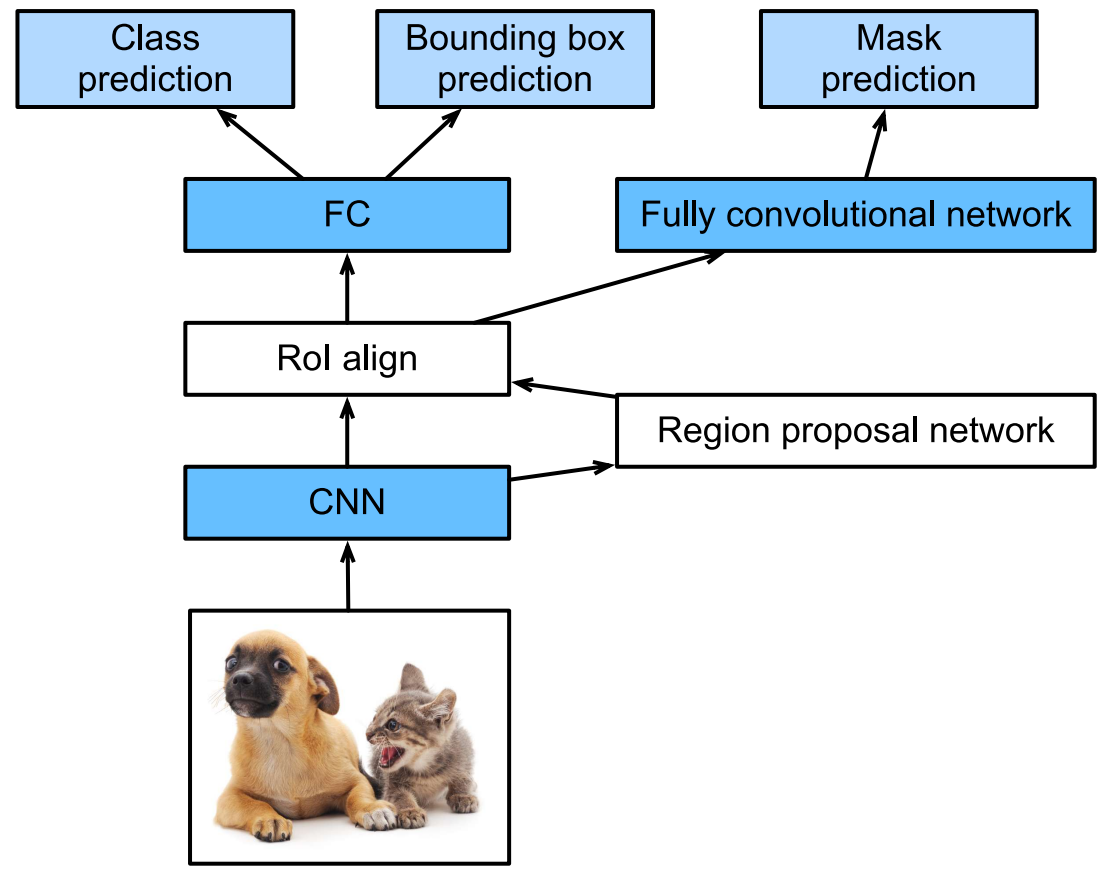

Fig. 14.8.5 The mask R-CNN model.

## 14.8.5 Summary

- The R-CNN extracts many region proposals from the input image, uses a CNN to perform forward propagation on each region proposal to extract its features, then uses these features to predict the class and bounding box of this region proposal.
- One of the major improvements of the fast R-CNN from the R-CNN is that the CNN forward propagation is only performed on the entire image. It also introduces the region of interest pooling layer, so that features of the same shape can be further extracted for regions of interest that have different shapes.
- The faster R-CNN replaces the selective search used in the fast R-CNN with a jointly trained region proposal network, so that the former can stay accurate in object detection with a reduced number of region proposals.
- Based on the faster R-CNN, the mask R-CNN additionally introduces a fully convolutional network, so as to leverage pixel-level labels to further improve the accuracy of object detection.

## 14.8.6 Exercises

1. Can we frame object detection as a single regression problem, such as predicting bounding boxes and class probabilities? You may refer to the design of the YOLO model (Redmon *et al.*, 2016).
2. Compare single shot multibox detection with the methods introduced in this section. What are their major differences? You may refer to Figure 2 of Zhao *et al.* (2019).

Discussions[219].

219

## 14.9 Semantic Segmentation and the Dataset

When discussing object detection tasks in Section 14.3–Section 14.8, rectangular bounding boxes are used to label and predict objects in images. This section will discuss the problem of *semantic segmentation*, which focuses on how to divide an image into regions belonging to different semantic classes. Different from object detection, semantic segmentation recognizes and understands what are in images in pixel level: its labeling and prediction of semantic regions are in pixel level. Fig. 14.9.1 shows the labels of the dog, cat, and background of the image in semantic segmentation. Compared with in object detection, the pixel-level borders labeled in semantic segmentation are obviously more fine-grained.

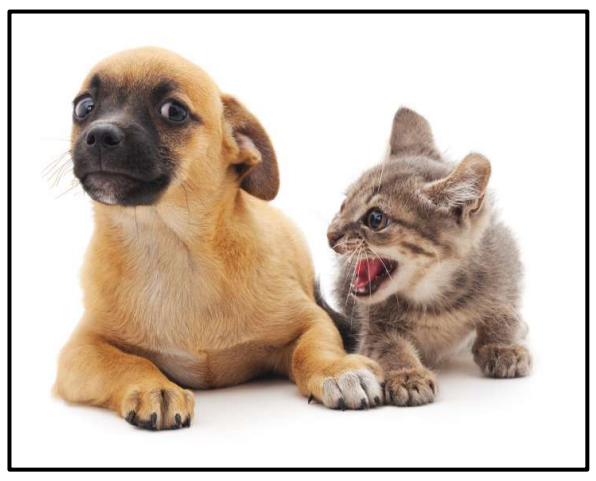

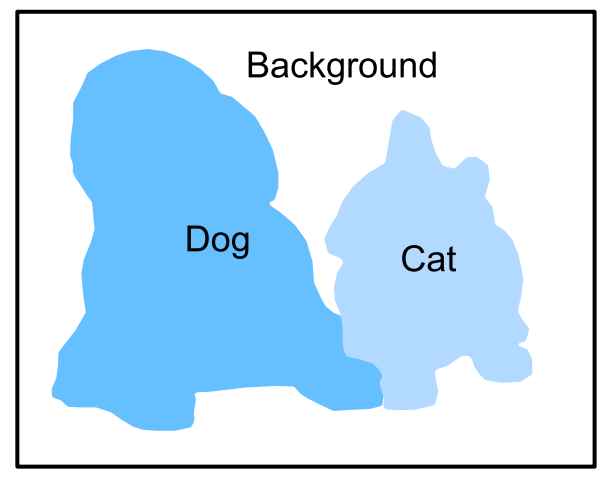

Fig. 14.9.1 Labels of the dog, cat, and background of the image in semantic segmentation.

### 14.9.1 Image Segmentation and Instance Segmentation

There are also two important tasks in the field of computer vision that are similar to semantic segmentation, namely image segmentation and instance segmentation. We will briefly distinguish them from semantic segmentation as follows.

- *Image segmentation* divides an image into several constituent regions. The methods for this type of problem usually make use of the correlation between pixels in the image. It does not need label information about image pixels during training, and it cannot guarantee that the segmented regions will have the semantics that we hope to obtain during prediction. Taking the image in Fig. 14.9.1 as input, image segmentation may divide the dog into two regions: one covers the mouth and eyes which are mainly black, and the other covers the rest of the body which is mainly yellow.

- *Instance segmentation* is also called *simultaneous detection and segmentation*. It studies how to recognize the pixel-level regions of each object instance in an image. Different from semantic segmentation, instance segmentation needs to distinguish not only semantics, but also different object instances. For example, if there are two dogs in the image, instance segmentation needs to distinguish which of the two dogs a pixel belongs to.

## 14.9.2 The Pascal VOC2012 Semantic Segmentation Dataset

On of the most important semantic segmentation dataset is Pascal VOC2012[220]. In the following, we will take a look at this dataset.

```
%matplotlib inline
import os
import torch
import torchvision
from d2l import torch as d2l
```

The tar file of the dataset is about 2 GB, so it may take a while to download the file. The extracted dataset is located at `../data/VOCdevkit/VOC2012`.

```
#@save
d2l.DATA_HUB['voc2012'] = (d2l.DATA_URL + 'VOCtrainval_11-May-2012.tar',
                           '4e443f8a2eca6b1dac8a6c57641b67dd40621a49')

voc_dir = d2l.download_extract('voc2012', 'VOCdevkit/VOC2012')
```

After entering the path `../data/VOCdevkit/VOC2012`, we can see the different components of the dataset. The `ImageSets/Segmentation` path contains text files that specify training and test samples, while the `JPEGImages` and `SegmentationClass` paths store the input image and label for each example, respectively. The label here is also in the image format, with the same size as its labeled input image. Besides, pixels with the same color in any label image belong to the same semantic class. The following defines the `read_voc_images` function to read all the input images and labels into the memory.

```
#@save
def read_voc_images(voc_dir, is_train=True):
    """Read all VOC feature and label images."""
    txt_fname = os.path.join(voc_dir, 'ImageSets', 'Segmentation',
                             'train.txt' if is_train else 'val.txt')
    mode = torchvision.io.image.ImageReadMode.RGB
    with open(txt_fname, 'r') as f:
        images = f.read().split()
    features, labels = [], []
    for i, fname in enumerate(images):
        features.append(torchvision.io.read_image(os.path.join(
```

```
            voc_dir, 'JPEGImages', f'{fname}.jpg')))
        labels.append(torchvision.io.read_image(os.path.join(
            voc_dir, 'SegmentationClass' ,f'{fname}.png'), mode))
    return features, labels

train_features, train_labels = read_voc_images(voc_dir, True)
```

We draw the first five input images and their labels. In the label images, white and black represent borders and background, respectively, while the other colors correspond to different classes.

```
n = 5
imgs = train_features[:n] + train_labels[:n]
imgs = [img.permute(1,2,0) for img in imgs]
d2l.show_images(imgs, 2, n);
```

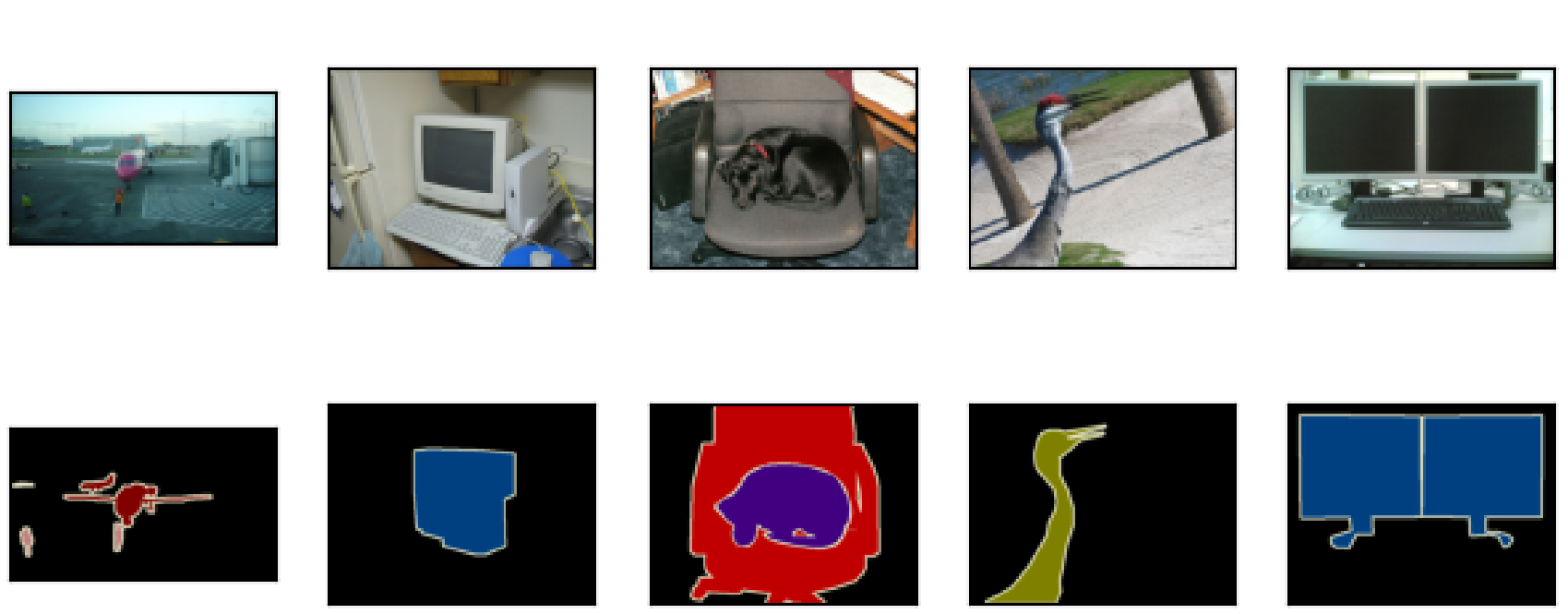

Next, we enumerate the RGB color values and class names for all the labels in this dataset.

```
#@save
VOC_COLORMAP = [[0, 0, 0], [128, 0, 0], [0, 128, 0], [128, 128, 0],
                [0, 0, 128], [128, 0, 128], [0, 128, 128], [128, 128, 128],
                [64, 0, 0], [192, 0, 0], [64, 128, 0], [192, 128, 0],
                [64, 0, 128], [192, 0, 128], [64, 128, 128], [192, 128, 128],
                [0, 64, 0], [128, 64, 0], [0, 192, 0], [128, 192, 0],
                [0, 64, 128]]

#@save
VOC_CLASSES = ['background', 'aeroplane', 'bicycle', 'bird', 'boat',
               'bottle', 'bus', 'car', 'cat', 'chair', 'cow',
               'diningtable', 'dog', 'horse', 'motorbike', 'person',
               'potted plant', 'sheep', 'sofa', 'train', 'tv/monitor']
```

With the two constants defined above, we can conveniently find the class index for each pixel in a label. We define the `voc_colormap2label` function to build the mapping from the above RGB color values to class indices, and the `voc_label_indices` function to map any RGB values to their class indices in this Pascal VOC2012 dataset.

```
#@save
def voc_colormap2label():
    """Build the mapping from RGB to class indices for VOC labels."""
    colormap2label = torch.zeros(256 ** 3, dtype=torch.long)
    for i, colormap in enumerate(VOC_COLORMAP):
        colormap2label[
            (colormap[0] * 256 + colormap[1]) * 256 + colormap[2]] = i
    return colormap2label

#@save
def voc_label_indices(colormap, colormap2label):
    """Map any RGB values in VOC labels to their class indices."""
    colormap = colormap.permute(1, 2, 0).numpy().astype('int32')
    idx = ((colormap[:, :, 0] * 256 + colormap[:, :, 1]) * 256
           + colormap[:, :, 2])
    return colormap2label[idx]
```

For example, in the first example image, the class index for the front part of the airplane is 1, while the background index is 0.

```
y = voc_label_indices(train_labels[0], voc_colormap2label())
y[105:115, 130:140], VOC_CLASSES[1]
```

```
(tensor([[0, 0, 0, 0, 0, 0, 0, 0, 0, 1],
         [0, 0, 0, 0, 0, 0, 0, 1, 1, 1],
         [0, 0, 0, 0, 0, 0, 1, 1, 1, 1],
         [0, 0, 0, 0, 0, 1, 1, 1, 1, 1],
         [0, 0, 0, 0, 0, 1, 1, 1, 1, 1],
         [0, 0, 0, 0, 1, 1, 1, 1, 1, 1],
         [0, 0, 0, 0, 0, 1, 1, 1, 1, 1],
         [0, 0, 0, 0, 0, 1, 1, 1, 1, 1],
         [0, 0, 0, 0, 0, 0, 1, 1, 1, 1],
         [0, 0, 0, 0, 0, 0, 0, 0, 1, 1]]),
 'aeroplane')
```

## Data Preprocessing

In previous experiments such as in Section 8.1–Section 8.4, images are rescaled to fit the model's required input shape. However, in semantic segmentation, doing so requires rescaling the predicted pixel classes back to the original shape of the input image. Such rescaling may be inaccurate, especially for segmented regions with different classes. To avoid this issue, we crop the image to a *fixed* shape instead of rescaling. Specifically, using random cropping from image augmentation, we crop the same area of the input image and the label.

```
#@save
def voc_rand_crop(feature, label, height, width):
    """Randomly crop both feature and label images."""
    rect = torchvision.transforms.RandomCrop.get_params(
```

```
        feature, (height, width))
    feature = torchvision.transforms.functional.crop(feature, *rect)
    label = torchvision.transforms.functional.crop(label, *rect)
    return feature, label
```

```
imgs = []
for _ in range(n):
    imgs += voc_rand_crop(train_features[0], train_labels[0], 200, 300)

imgs = [img.permute(1, 2, 0) for img in imgs]
d2l.show_images(imgs[::2] + imgs[1::2], 2, n);
```

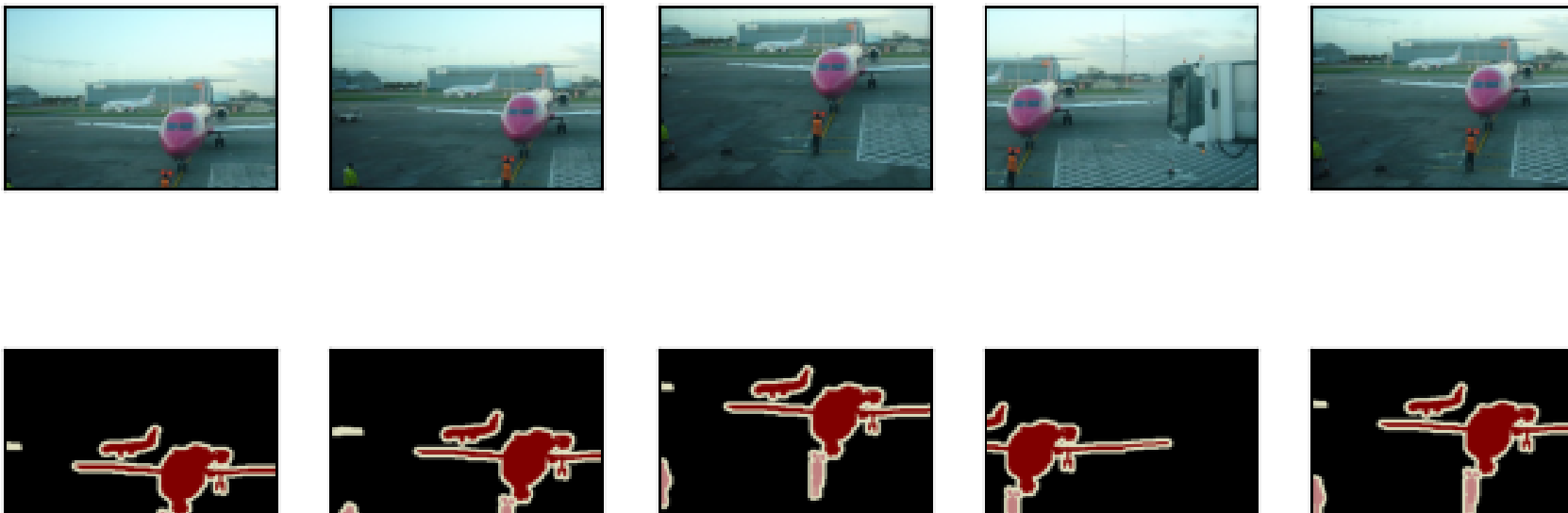

## Custom Semantic Segmentation Dataset Class

We define a custom semantic segmentation dataset class `VOCSegDataset` by inheriting the `Dataset` class provided by high-level APIs. By implementing the `__getitem__` function, we can arbitrarily access the input image indexed as `idx` in the dataset and the class index of each pixel in this image. Since some images in the dataset have a smaller size than the output size of random cropping, these examples are filtered out by a custom `filter` function. In addition, we also define the `normalize_image` function to standardize the values of the three RGB channels of input images.

```
#@save
class VOCSegDataset(torch.utils.data.Dataset):
    """A customized dataset to load the VOC dataset."""

    def __init__(self, is_train, crop_size, voc_dir):
        self.transform = torchvision.transforms.Normalize(
            mean=[0.485, 0.456, 0.406], std=[0.229, 0.224, 0.225])
        self.crop_size = crop_size
        features, labels = read_voc_images(voc_dir, is_train=is_train)
        self.features = [self.normalize_image(feature)
                         for feature in self.filter(features)]
        self.labels = self.filter(labels)
        self.colormap2label = voc_colormap2label()
```

```
        print('read ' + str(len(self.features)) + ' examples')

    def normalize_image(self, img):
        return self.transform(img.float() / 255)

    def filter(self, imgs):
        return [img for img in imgs if (
            img.shape[1] >= self.crop_size[0] and
            img.shape[2] >= self.crop_size[1])]

    def __getitem__(self, idx):
        feature, label = voc_rand_crop(self.features[idx], self.labels[idx],
                                       *self.crop_size)
        return (feature, voc_label_indices(label, self.colormap2label))

    def __len__(self):
        return len(self.features)
```

### Reading the Dataset

We use the custom `VOCSegDataset` class to create instances of the training set and test set, respectively. Suppose that we specify that the output shape of randomly cropped images is $320 \times 480$. Below we can view the number of examples that are retained in the training set and test set.

```
crop_size = (320, 480)
voc_train = VOCSegDataset(True, crop_size, voc_dir)
voc_test = VOCSegDataset(False, crop_size, voc_dir)
```

```
read 1114 examples
read 1078 examples
```

Setting the batch size to 64, we define the data iterator for the training set. Let's print the shape of the first minibatch. Different from in image classification or object detection, labels here are three-dimensional tensors.

```
batch_size = 64
train_iter = torch.utils.data.DataLoader(voc_train, batch_size, shuffle=True,
                                    drop_last=True,
                                    num_workers=d2l.get_dataloader_workers())
for X, Y in train_iter:
    print(X.shape)
    print(Y.shape)
    break
```

```
torch.Size([64, 3, 320, 480])
torch.Size([64, 320, 480])
```

### Putting It All Together

Finally, we define the following `load_data_voc` function to download and read the Pascal VOC2012 semantic segmentation dataset. It returns data iterators for both the training and test datasets.

```
#@save
def load_data_voc(batch_size, crop_size):
    """Load the VOC semantic segmentation dataset."""
    voc_dir = d2l.download_extract('voc2012', os.path.join(
        'VOCdevkit', 'VOC2012'))
    num_workers = d2l.get_dataloader_workers()
    train_iter = torch.utils.data.DataLoader(
        VOCSegDataset(True, crop_size, voc_dir), batch_size,
        shuffle=True, drop_last=True, num_workers=num_workers)
    test_iter = torch.utils.data.DataLoader(
        VOCSegDataset(False, crop_size, voc_dir), batch_size,
        drop_last=True, num_workers=num_workers)
    return train_iter, test_iter
```

## 14.9.3 Summary

- Semantic segmentation recognizes and understands what are in an image in pixel level by dividing the image into regions belonging to different semantic classes.
- One of the most important semantic segmentation dataset is Pascal VOC2012.
- In semantic segmentation, since the input image and label correspond one-to-one on the pixel, the input image is randomly cropped to a fixed shape rather than rescaled.

## 14.9.4 Exercises

1. How can semantic segmentation be applied in autonomous vehicles and medical image diagnostics? Can you think of other applications?
2. Recall the descriptions of data augmentation in Section 14.1. Which of the image augmentation methods used in image classification would be infeasible to be applied in semantic segmentation?

Discussions[221].

# 14.10 Transposed Convolution

The CNN layers we have seen so far, such as convolutional layers (Section 7.2) and pooling layers (Section 7.5), typically reduce (downsample) the spatial dimensions (height and width) of the input, or keep them unchanged. In semantic segmentation that classifies at pixel-level, it will be convenient if the spatial dimensions of the input and output are the same. For example, the channel dimension at one output pixel can hold the classification results for the input pixel at the same spatial position.

To achieve this, especially after the spatial dimensions are reduced by CNN layers, we can use another type of CNN layers that can increase (upsample) the spatial dimensions of intermediate feature maps. In this section, we will introduce *transposed convolution*, which is also called *fractionally-strided convolution* (Dumoulin and Visin, 2016), for reversing downsampling operations by the convolution.

```
import torch
from torch import nn
from d2l import torch as d2l
```

## 14.10.1 Basic Operation

Ignoring channels for now, let's begin with the basic transposed convolution operation with stride of 1 and no padding. Suppose that we are given a $n_h \times n_w$ input tensor and a $k_h \times k_w$ kernel. Sliding the kernel window with stride of 1 for $n_w$ times in each row and $n_h$ times in each column yields a total of $n_h n_w$ intermediate results. Each intermediate result is a $(n_h + k_h - 1) \times (n_w + k_w - 1)$ tensor that are initialized as zeros. To compute each intermediate tensor, each element in the input tensor is multiplied by the kernel so that the resulting $k_h \times k_w$ tensor replaces a portion in each intermediate tensor. Note that the position of the replaced portion in each intermediate tensor corresponds to the position of the element in the input tensor used for the computation. In the end, all the intermediate results are summed over to produce the output.

As an example, Fig. 14.10.1 illustrates how transposed convolution with a $2 \times 2$ kernel is computed for a $2 \times 2$ input tensor.

We can implement this basic transposed convolution operation `trans_conv` for a input matrix `X` and a kernel matrix `K`.

```
def trans_conv(X, K):
    h, w = K.shape
    Y = torch.zeros((X.shape[0] + h - 1, X.shape[1] + w - 1))
    for i in range(X.shape[0]):
        for j in range(X.shape[1]):
```

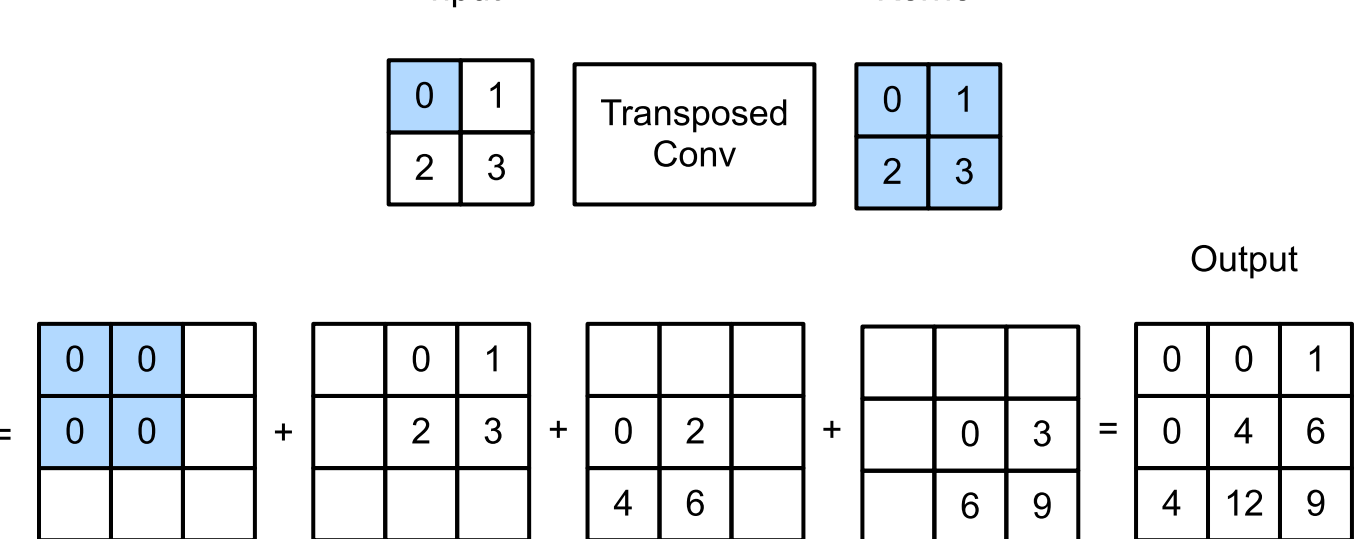

Fig. 14.10.1 Transposed convolution with a $2 \times 2$ kernel. The shaded portions are a portion of an intermediate tensor as well as the input and kernel tensor elements used for the computation.

```
            Y[i: i + h, j: j + w] += X[i, j] * K
    return Y
```

In contrast to the regular convolution (in Section 7.2) that *reduces* input elements via the kernel, the transposed convolution *broadcasts* input elements via the kernel, thereby producing an output that is larger than the input. We can construct the input tensor X and the kernel tensor K from Fig. 14.10.1 to validate the output of the above implementation of the basic two-dimensional transposed convolution operation.

```
X = torch.tensor([[0.0, 1.0], [2.0, 3.0]])
K = torch.tensor([[0.0, 1.0], [2.0, 3.0]])
trans_conv(X, K)
```

```
tensor([[ 0.,  0.,  1.],
        [ 0.,  4.,  6.],
        [ 4., 12.,  9.]])
```

Alternatively, when the input X and kernel K are both four-dimensional tensors, we can use high-level APIs to obtain the same results.

```
X, K = X.reshape(1, 1, 2, 2), K.reshape(1, 1, 2, 2)
tconv = nn.ConvTranspose2d(1, 1, kernel_size=2, bias=False)
tconv.weight.data = K
tconv(X)
```

```
tensor([[[[ 0.,  0.,  1.],
          [ 0.,  4.,  6.],
          [ 4., 12.,  9.]]]], grad_fn=<ConvolutionBackward0>)
```

## 14.10.2 Padding, Strides, and Multiple Channels

---

Different from in the regular convolution where padding is applied to input, it is applied to output in the transposed convolution. For example, when specifying the padding number on either side of the height and width as 1, the first and last rows and columns will be removed from the transposed convolution output.

```
tconv = nn.ConvTranspose2d(1, 1, kernel_size=2, padding=1, bias=False)
tconv.weight.data = K
tconv(X)
```

```
tensor([[[[4.]]]], grad_fn=<ConvolutionBackward0>)
```

In the transposed convolution, strides are specified for intermediate results (thus output), not for input. Using the same input and kernel tensors from Fig. 14.10.1, changing the stride from 1 to 2 increases both the height and weight of intermediate tensors, hence the output tensor in Fig. 14.10.2.

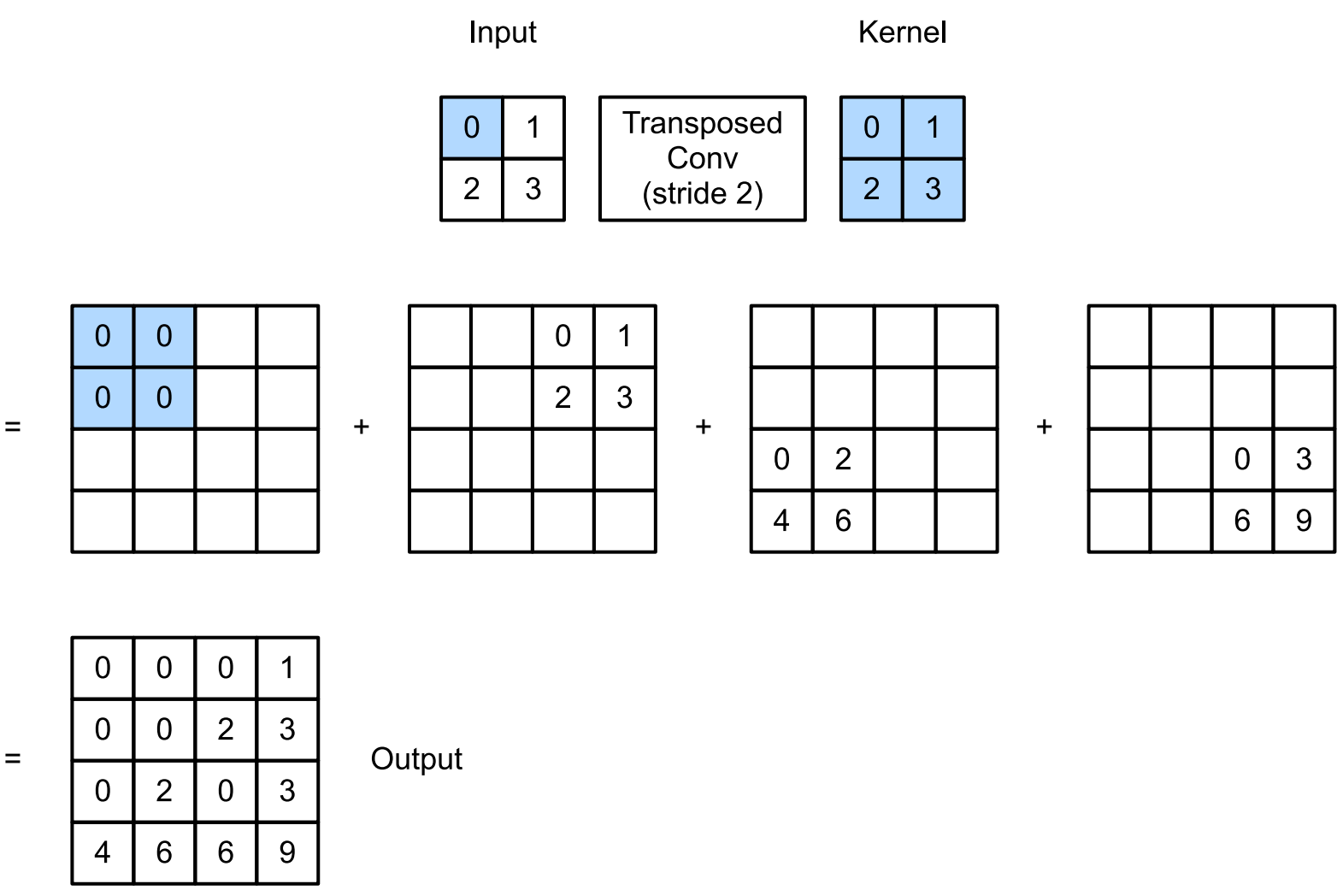

Fig. 14.10.2 Transposed convolution with a $2 \times 2$ kernel with stride of 2. The shaded portions are a portion of an intermediate tensor as well as the input and kernel tensor elements used for the computation.

The following code snippet can validate the transposed convolution output for stride of 2 in Fig. 14.10.2.

```
tconv = nn.ConvTranspose2d(1, 1, kernel_size=2, stride=2, bias=False)
tconv.weight.data = K
tconv(X)
```

```
tensor([[[[0., 0., 0., 1.],
          [0., 0., 2., 3.],
          [0., 2., 0., 3.],
          [4., 6., 6., 9.]]]], grad_fn=<ConvolutionBackward0>)
```

For multiple input and output channels, the transposed convolution works in the same way as the regular convolution. Suppose that the input has $c_i$ channels, and that the transposed convolution assigns a $k_h \times k_w$ kernel tensor to each input channel. When multiple output channels are specified, we will have a $c_i \times k_h \times k_w$ kernel for each output channel.

As in all, if we feed $\mathsf{X}$ into a convolutional layer $f$ to output $\mathsf{Y} = f(\mathsf{X})$ and create a transposed convolutional layer $g$ with the same hyperparameters as $f$ except for the number of output channels being the number of channels in $\mathsf{X}$, then $g(Y)$ will have the same shape as $\mathsf{X}$. This can be illustrated in the following example.

```
X = torch.rand(size=(1, 10, 16, 16))
conv = nn.Conv2d(10, 20, kernel_size=5, padding=2, stride=3)
tconv = nn.ConvTranspose2d(20, 10, kernel_size=5, padding=2, stride=3)
tconv(conv(X)).shape == X.shape
```

```
True
```

## 14.10.3 Connection to Matrix Transposition

The transposed convolution is named after the matrix transposition. To explain, let's first see how to implement convolutions using matrix multiplications. In the example below, we define a $3 \times 3$ input `X` and a $2 \times 2$ convolution kernel `K`, and then use the `corr2d` function to compute the convolution output `Y`.

```
X = torch.arange(9.0).reshape(3, 3)
K = torch.tensor([[1.0, 2.0], [3.0, 4.0]])
Y = d2l.corr2d(X, K)
Y
```

```
tensor([[27., 37.],
        [57., 67.]])
```

Next, we rewrite the convolution kernel `K` as a sparse weight matrix `W` containing a lot of zeros. The shape of the weight matrix is (4, 9), where the non-zero elements come from the convolution kernel `K`.

```
def kernel2matrix(K):
    k, W = torch.zeros(5), torch.zeros((4, 9))
    k[:2], k[3:5] = K[0, :], K[1, :]
```

```
    W[0, :5], W[1, 1:6], W[2, 3:8], W[3, 4:] = k, k, k, k
    return W

W = kernel2matrix(K)
W
```

```
tensor([[1., 2., 0., 3., 4., 0., 0., 0., 0.],
        [0., 1., 2., 0., 3., 4., 0., 0., 0.],
        [0., 0., 0., 1., 2., 0., 3., 4., 0.],
        [0., 0., 0., 0., 1., 2., 0., 3., 4.]])
```

Concatenate the input `X` row by row to get a vector of length 9. Then the matrix multiplication of `W` and the vectorized `X` gives a vector of length 4. After reshaping it, we can obtain the same result `Y` from the original convolution operation above: we just implemented convolutions using matrix multiplications.

```
Y == torch.matmul(W, X.reshape(-1)).reshape(2, 2)
```

```
tensor([[True, True],
        [True, True]])
```

Likewise, we can implement transposed convolutions using matrix multiplications. In the following example, we take the $2 \times 2$ output `Y` from the above regular convolution as input to the transposed convolution. To implement this operation by multiplying matrices, we only need to transpose the weight matrix `W` with the new shape $(9, 4)$.

```
Z = trans_conv(Y, K)
Z == torch.matmul(W.T, Y.reshape(-1)).reshape(3, 3)
```

```
tensor([[True, True, True],
        [True, True, True],
        [True, True, True]])
```

Consider implementing the convolution by multiplying matrices. Given an input vector $\mathbf{x}$ and a weight matrix $\mathbf{W}$, the forward propagation function of the convolution can be implemented by multiplying its input with the weight matrix and outputting a vector $\mathbf{y} = \mathbf{W}\mathbf{x}$. Since backpropagation follows the chain rule and $\nabla_{\mathbf{x}}\mathbf{y} = \mathbf{W}^\top$, the backpropagation function of the convolution can be implemented by multiplying its input with the transposed weight matrix $\mathbf{W}^\top$. Therefore, the transposed convolutional layer can just exchange the forward propagation function and the backpropagation function of the convolutional layer: its forward propagation and backpropagation functions multiply their input vector with $\mathbf{W}^\top$ and $\mathbf{W}$, respectively.

### 14.10.4 Summary

- In contrast to the regular convolution that reduces input elements via the kernel, the transposed convolution broadcasts input elements via the kernel, thereby producing an output that is larger than the input.
- If we feed $\mathsf{X}$ into a convolutional layer $f$ to output $\mathsf{Y} = f(\mathsf{X})$ and create a transposed convolutional layer $g$ with the same hyperparameters as $f$ except for the number of output channels being the number of channels in $\mathsf{X}$, then $g(Y)$ will have the same shape as $\mathsf{X}$.
- We can implement convolutions using matrix multiplications. The transposed convolutional layer can just exchange the forward propagation function and the backpropagation function of the convolutional layer.

### 14.10.5 Exercises

1. In Section 14.10.3, the convolution input `X` and the transposed convolution output `Z` have the same shape. Do they have the same value? Why?
2. Is it efficient to use matrix multiplications to implement convolutions? Why?

Discussions[222].

## 14.11 Fully Convolutional Networks

As discussed in Section 14.9, semantic segmentation classifies images in pixel level. A fully convolutional network (FCN) uses a convolutional neural network to transform image pixels to pixel classes (Long *et al.*, 2015). Unlike the CNNs that we encountered earlier for image classification or object detection, a fully convolutional network transforms the height and width of intermediate feature maps back to those of the input image: this is achieved by the transposed convolutional layer introduced in Section 14.10. As a result, the classification output and the input image have a one-to-one correspondence in pixel level: the channel dimension at any output pixel holds the classification results for the input pixel at the same spatial position.

```
%matplotlib inline
import torch
import torchvision
from torch import nn
from torch.nn import functional as F
from d2l import torch as d2l
```

### 14.11.1 The Model

Here we describe the basic design of the fully convolutional network model. As shown in Fig. 14.11.1, this model first uses a CNN to extract image features, then transforms the number of channels into the number of classes via a $1 \times 1$ convolutional layer, and finally transforms the height and width of the feature maps to those of the input image via the transposed convolution introduced in Section 14.10. As a result, the model output has the same height and width as the input image, where the output channel contains the predicted classes for the input pixel at the same spatial position.

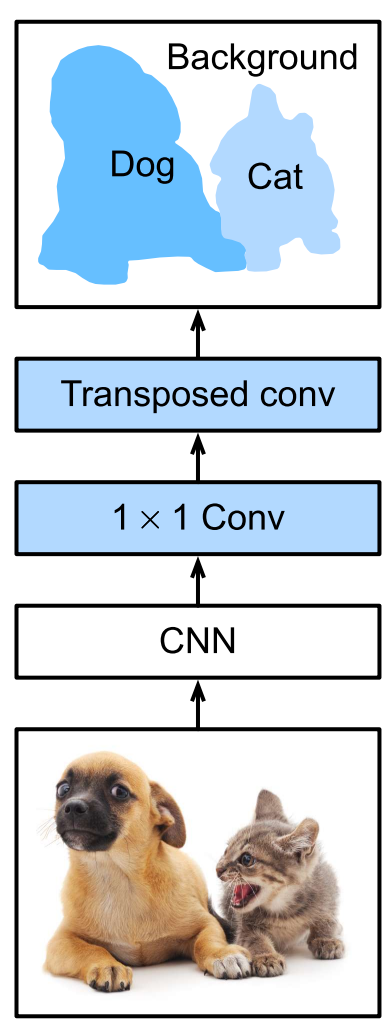

Fig. 14.11.1 Fully convolutional network.

Below, we use a ResNet-18 model pretrained on the ImageNet dataset to extract image features and denote the model instance as `pretrained_net`. The last few layers of this model include a global average pooling layer and a fully connected layer: they are not needed in the fully convolutional network.

```
pretrained_net = torchvision.models.resnet18(pretrained=True)
list(pretrained_net.children())[-3:]
```

```
[Sequential(
   (0): BasicBlock(
     (conv1): Conv2d(256, 512, kernel_size=(3, 3), stride=(2, 2), padding=(1,␣
↪1), bias=False)
     (bn1): BatchNorm2d(512, eps=1e-05, momentum=0.1, affine=True, track_
↪running_stats=True)
     (relu): ReLU(inplace=True)
     (conv2): Conv2d(512, 512, kernel_size=(3, 3), stride=(1, 1), padding=(1,␣
↪1), bias=False)
     (bn2): BatchNorm2d(512, eps=1e-05, momentum=0.1, affine=True, track_
↪running_stats=True)
```

```
     (downsample): Sequential(
       (0): Conv2d(256, 512, kernel_size=(1, 1), stride=(2, 2), bias=False)
       (1): BatchNorm2d(512, eps=1e-05, momentum=0.1, affine=True, track_
↪running_stats=True)
     )
   )
   (1): BasicBlock(
     (conv1): Conv2d(512, 512, kernel_size=(3, 3), stride=(1, 1), padding=(1,␣
↪1), bias=False)
     (bn1): BatchNorm2d(512, eps=1e-05, momentum=0.1, affine=True, track_
↪running_stats=True)
     (relu): ReLU(inplace=True)
     (conv2): Conv2d(512, 512, kernel_size=(3, 3), stride=(1, 1), padding=(1,␣
↪1), bias=False)
     (bn2): BatchNorm2d(512, eps=1e-05, momentum=0.1, affine=True, track_
↪running_stats=True)
   )
 ),
 AdaptiveAvgPool2d(output_size=(1, 1)),
 Linear(in_features=512, out_features=1000, bias=True)]
```

Next, we create the fully convolutional network instance `net`. It copies all the pretrained layers in the ResNet-18 except for the final global average pooling layer and the fully connected layer that are closest to the output.

```
net = nn.Sequential(*list(pretrained_net.children())[:-2])
```

Given an input with height and width of 320 and 480 respectively, the forward propagation of `net` reduces the input height and width to 1/32 of the original, namely 10 and 15.

```
X = torch.rand(size=(1, 3, 320, 480))
net(X).shape
```

```
torch.Size([1, 512, 10, 15])
```

Next, we use a $1\times 1$ convolutional layer to transform the number of output channels into the number of classes (21) of the Pascal VOC2012 dataset. Finally, we need to increase the height and width of the feature maps by 32 times to change them back to the height and width of the input image. Recall how to calculate the output shape of a convolutional layer in Section 7.3. Since $(320-64+16\times2+32)/32 = 10$ and $(480-64+16\times2+32)/32 = 15$, we construct a transposed convolutional layer with stride of 32, setting the height and width of the kernel to $64$, the padding to $16$. In general, we can see that for stride $s$, padding $s/2$ (assuming $s/2$ is an integer), and the height and width of the kernel $2s$, the transposed convolution will increase the height and width of the input by $s$ times.

```
num_classes = 21
net.add_module('final_conv', nn.Conv2d(512, num_classes, kernel_size=1))
```

```
net.add_module('transpose_conv', nn.ConvTranspose2d(num_classes, num_classes,
                                    kernel_size=64, padding=16, stride=32))
```

## 14.11.2 Initializing Transposed Convolutional Layers

We already know that transposed convolutional layers can increase the height and width of feature maps. In image processing, we may need to scale up an image, i.e., *upsampling*. *Bilinear interpolation* is one of the commonly used upsampling techniques. It is also often used for initializing transposed convolutional layers.

To explain bilinear interpolation, say that given an input image we want to calculate each pixel of the upsampled output image. In order to calculate the pixel of the output image at coordinate $(x, y)$, first map $(x, y)$ to coordinate $(x', y')$ on the input image, for example, according to the ratio of the input size to the output size. Note that the mapped $x'$ and $y'$ are real numbers. Then, find the four pixels closest to coordinate $(x', y')$ on the input image. Finally, the pixel of the output image at coordinate $(x, y)$ is calculated based on these four closest pixels on the input image and their relative distance from $(x', y')$.

Upsampling of bilinear interpolation can be implemented by the transposed convolutional layer with the kernel constructed by the following `bilinear_kernel` function. Due to space limitations, we only provide the implementation of the `bilinear_kernel` function below without discussions on its algorithm design.

```
def bilinear_kernel(in_channels, out_channels, kernel_size):
    factor = (kernel_size + 1) // 2
    if kernel_size % 2 == 1:
        center = factor - 1
    else:
        center = factor - 0.5
    og = (torch.arange(kernel_size).reshape(-1, 1),
          torch.arange(kernel_size).reshape(1, -1))
    filt = (1 - torch.abs(og[0] - center) / factor) * \
           (1 - torch.abs(og[1] - center) / factor)
    weight = torch.zeros((in_channels, out_channels,
                          kernel_size, kernel_size))
    weight[range(in_channels), range(out_channels), :, :] = filt
    return weight
```

Let's experiment with upsampling of bilinear interpolation that is implemented by a transposed convolutional layer. We construct a transposed convolutional layer that doubles the height and weight, and initialize its kernel with the `bilinear_kernel` function.

```
conv_trans = nn.ConvTranspose2d(3, 3, kernel_size=4, padding=1, stride=2,
                                bias=False)
conv_trans.weight.data.copy_(bilinear_kernel(3, 3, 4));
```

Read the image X and assign the upsampling output to Y. In order to print the image, we need to adjust the position of the channel dimension.

```
img = torchvision.transforms.ToTensor()(d2l.Image.open('../img/catdog.jpg'))
X = img.unsqueeze(0)
Y = conv_trans(X)
out_img = Y[0].permute(1, 2, 0).detach()
```

As we can see, the transposed convolutional layer increases both the height and width of the image by a factor of two. Except for the different scales in coordinates, the image scaled up by bilinear interpolation and the original image printed in Section 14.3 look the same.

```
d2l.set_figsize()
print('input image shape:', img.permute(1, 2, 0).shape)
d2l.plt.imshow(img.permute(1, 2, 0));
print('output image shape:', out_img.shape)
d2l.plt.imshow(out_img);
```

```
input image shape: torch.Size([561, 728, 3])
output image shape: torch.Size([1122, 1456, 3])
```

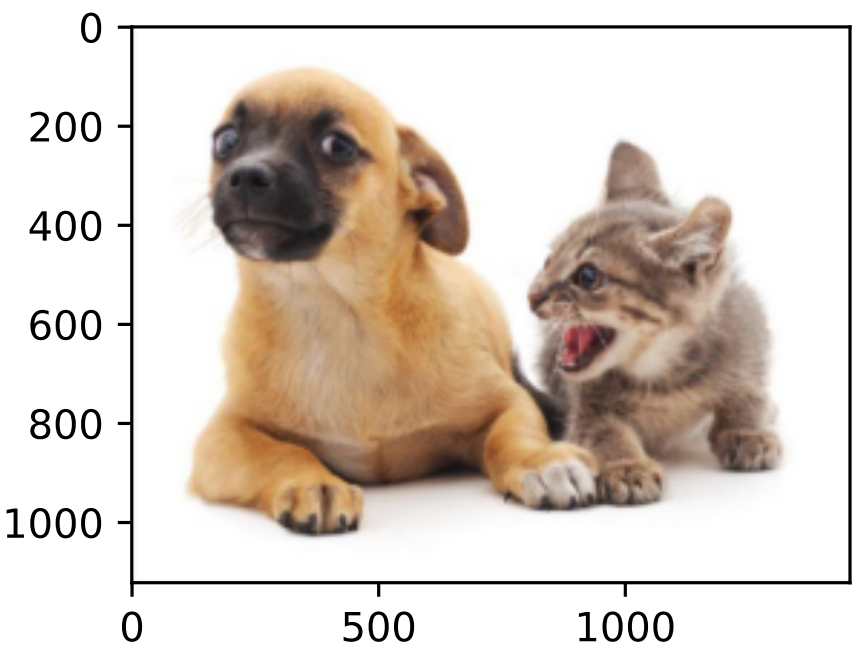

In a fully convolutional network, we initialize the transposed convolutional layer with upsampling of bilinear interpolation. For the $1 \times 1$ convolutional layer, we use Xavier initialization.

```
W = bilinear_kernel(num_classes, num_classes, 64)
net.transpose_conv.weight.data.copy_(W);
```

### 14.11.3 Reading the Dataset

We read the semantic segmentation dataset as introduced in Section 14.9. The output image shape of random cropping is specified as $320 \times 480$: both the height and width are divisible by 32.

```
batch_size, crop_size = 32, (320, 480)
train_iter, test_iter = d2l.load_data_voc(batch_size, crop_size)
```

```
read 1114 examples
read 1078 examples
```

## 14.11.4 Training

Now we can train our constructed fully convolutional network. The loss function and accuracy calculation here are not essentially different from those in image classification of earlier chapters. Because we use the output channel of the transposed convolutional layer to predict the class for each pixel, the channel dimension is specified in the loss calculation. In addition, the accuracy is calculated based on correctness of the predicted class for all the pixels.

```
def loss(inputs, targets):
    return F.cross_entropy(inputs, targets, reduction='none').mean(1).mean(1)

num_epochs, lr, wd, devices = 5, 0.001, 1e-3, d2l.try_all_gpus()
trainer = torch.optim.SGD(net.parameters(), lr=lr, weight_decay=wd)
d2l.train_ch13(net, train_iter, test_iter, loss, trainer, num_epochs, devices)
```

```
loss 0.448, train acc 0.862, test acc 0.852
195.7 examples/sec on [device(type='cuda', index=0), device(type='cuda',␣
↪index=1)]
```

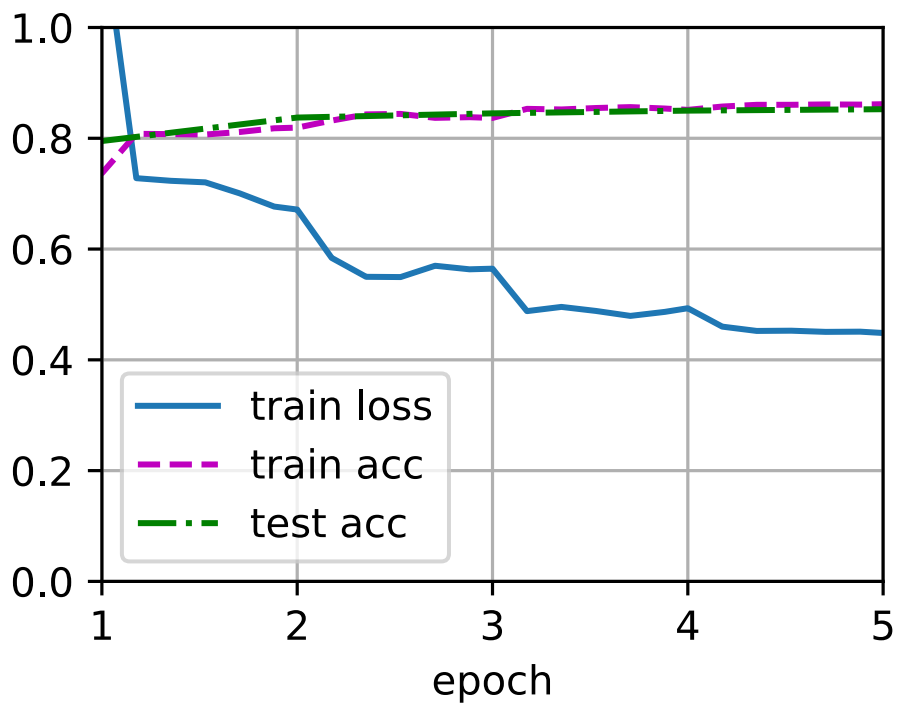

## 14.11.5 Prediction

When predicting, we need to standardize the input image in each channel and transform the image into the four-dimensional input format required by the CNN.

```
def predict(img):
    X = test_iter.dataset.normalize_image(img).unsqueeze(0)
    pred = net(X.to(devices[0])).argmax(dim=1)
    return pred.reshape(pred.shape[1], pred.shape[2])
```

To visualize the predicted class of each pixel, we map the predicted class back to its label color in the dataset.

```
def label2image(pred):
    colormap = torch.tensor(d2l.VOC_COLORMAP, device=devices[0])
    X = pred.long()
    return colormap[X, :]
```

Images in the test dataset vary in size and shape. Since the model uses a transposed convolutional layer with stride of 32, when the height or width of an input image is indivisible by 32, the output height or width of the transposed convolutional layer will deviate from the shape of the input image. In order to address this issue, we can crop multiple rectangular areas with height and width that are integer multiples of 32 in the image, and perform forward propagation on the pixels in these areas separately. Note that the union of these rectangular areas needs to completely cover the input image. When a pixel is covered by multiple rectangular areas, the average of the transposed convolution outputs in separate areas for this same pixel can be input to the softmax operation to predict the class.

For simplicity, we only read a few larger test images, and crop a $320 \times 480$ area for prediction starting from the upper-left corner of an image. For these test images, we print their cropped areas, prediction results, and ground-truth row by row.

```
voc_dir = d2l.download_extract('voc2012', 'VOCdevkit/VOC2012')
test_images, test_labels = d2l.read_voc_images(voc_dir, False)
n, imgs = 4, []
for i in range(n):
    crop_rect = (0, 0, 320, 480)
    X = torchvision.transforms.functional.crop(test_images[i], *crop_rect)
    pred = label2image(predict(X))
    imgs += [X.permute(1,2,0), pred.cpu(),
             torchvision.transforms.functional.crop(
                 test_labels[i], *crop_rect).permute(1,2,0)]
d2l.show_images(imgs[::3] + imgs[1::3] + imgs[2::3], 3, n, scale=2);
```

### 14.11.6 Summary

- The fully convolutional network first uses a CNN to extract image features, then transforms the number of channels into the number of classes via a $1 \times 1$ convolutional layer, and finally transforms the height and width of the feature maps to those of the input image via the transposed convolution.
- In a fully convolutional network, we can use upsampling of bilinear interpolation to initialize the transposed convolutional layer.

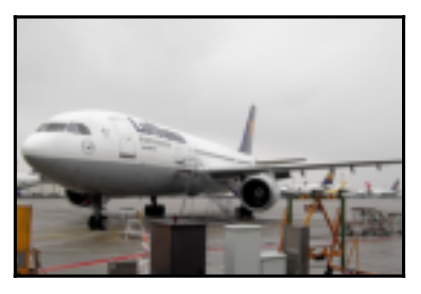
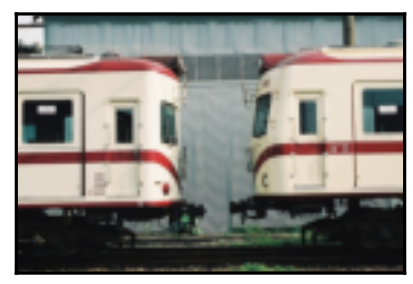
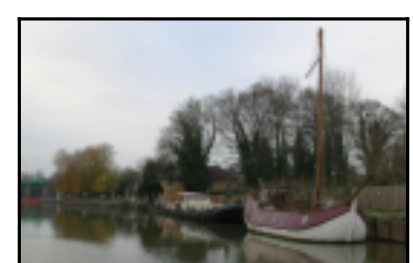
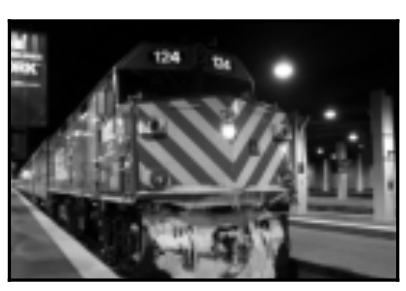
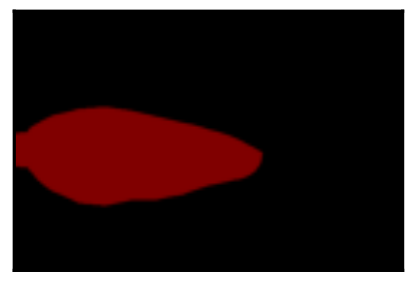
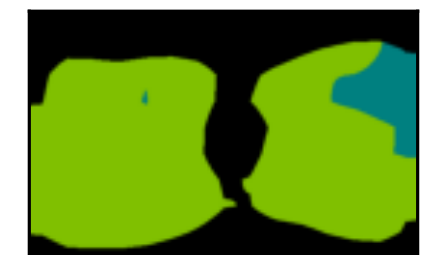
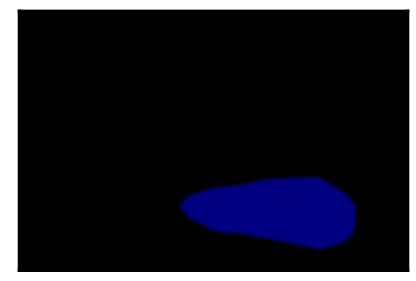
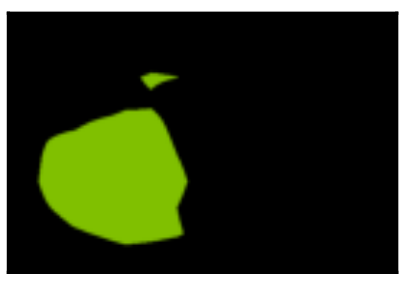
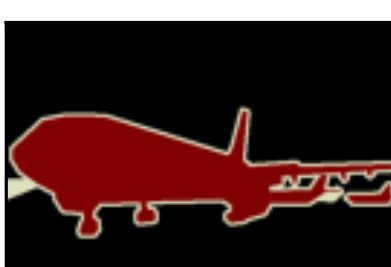
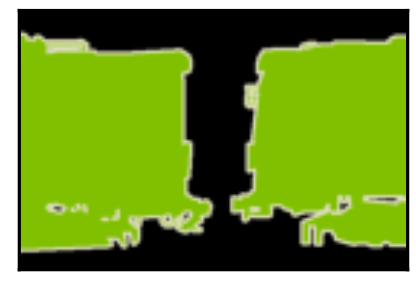
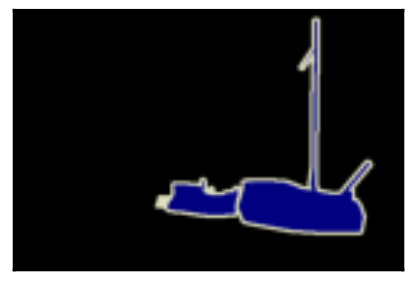
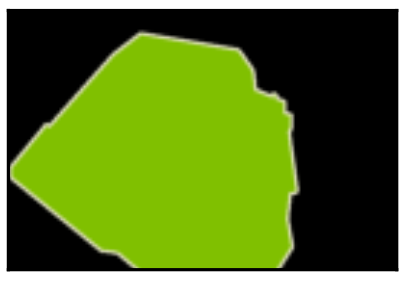

### 14.11.7 Exercises

1. If we use Xavier initialization for the transposed convolutional layer in the experiment, how does the result change?
2. Can you further improve the accuracy of the model by tuning the hyperparameters?
3. Predict the classes of all pixels in test images.
4. The original fully convolutional network paper also uses outputs of some intermediate CNN layers (Long *et al.*, 2015). Try to implement this idea.

Discussions[223].

## 14.12 Neural Style Transfer

If you are a photography enthusiast, you may be familiar with the filter. It can change the color style of photos so that landscape photos become sharper or portrait photos have whitened skins. However, one filter usually only changes one aspect of the photo. To apply an ideal style to a photo, you probably need to try many different filter combinations. This process is as complex as tuning the hyperparameters of a model.

In this section, we will leverage layerwise representations of a CNN to automatically apply the style of one image to another image, i.e., *style transfer* (Gatys *et al.*, 2016). This task needs two input images: one is the *content image* and the other is the *style image*. We will use neural networks to modify the content image to make it close to the style image in style. For example, the content image in Fig. 14.12.1 is a landscape photo taken by us in Mount Rainier National Park in the suburbs of Seattle, while the style image is an oil painting with the theme of autumn oak trees. In the output synthesized image, the oil brush strokes of the style image are applied, leading to more vivid colors, while preserving the main shape of the objects in the content image.

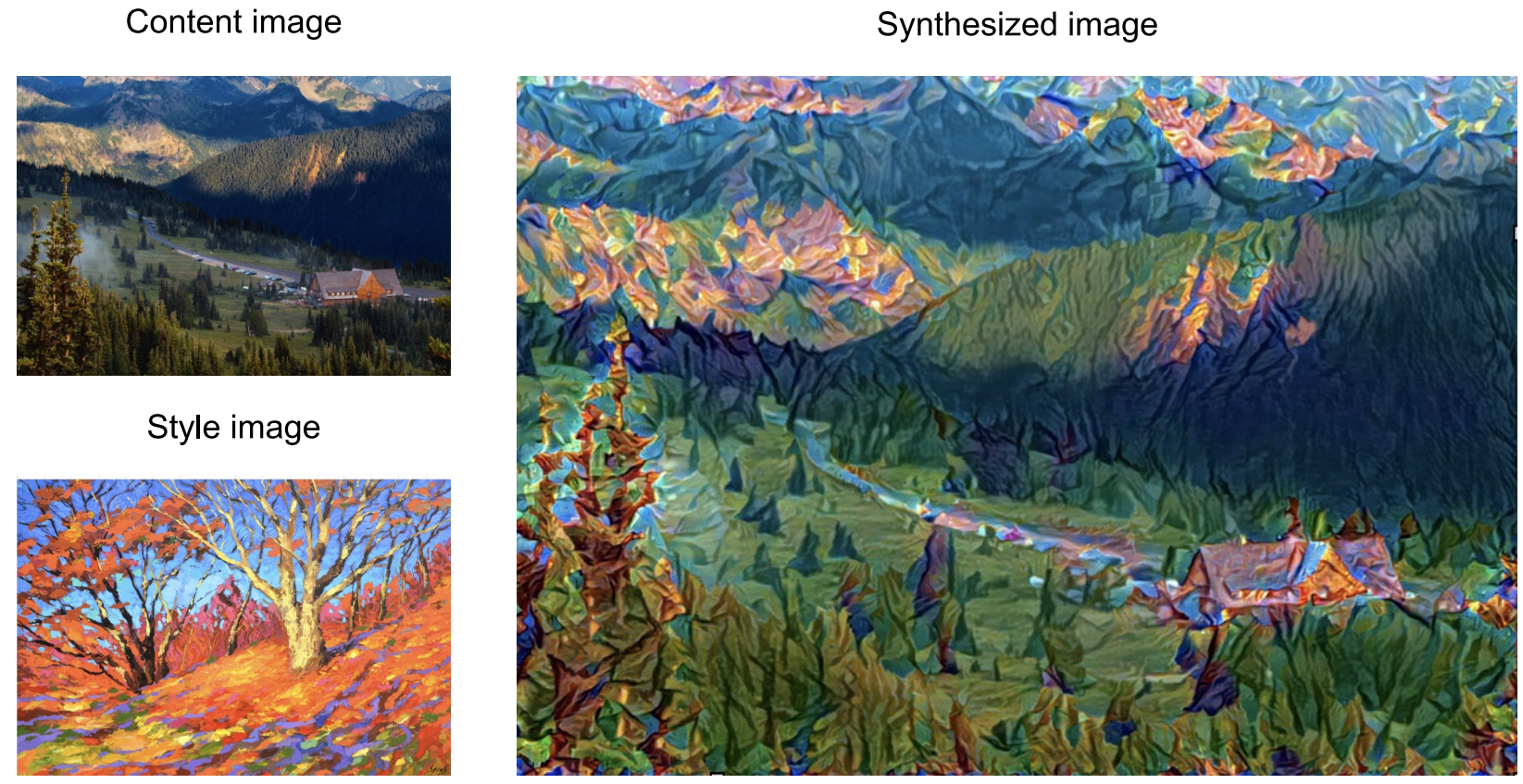

Fig. 14.12.1 Given content and style images, style transfer outputs a synthesized image.

## 14.12.1 Method

Fig. 14.12.2 illustrates the CNN-based style transfer method with a simplified example. First, we initialize the synthesized image, for example, into the content image. This synthesized image is the only variable that needs to be updated during the style transfer process, i.e., the model parameters to be updated during training. Then we choose a pretrained CNN to extract image features and freeze its model parameters during training. This deep CNN uses multiple layers to extract hierarchical features for images. We can choose the output of some of these layers as content features or style features. Take Fig. 14.12.2 as an example. The pretrained neural network here has 3 convolutional layers, where the second layer outputs the content features, and the first and third layers output the style features.

Next, we calculate the loss function of style transfer through forward propagation (direction of solid arrows), and update the model parameters (the synthesized image for output) through backpropagation (direction of dashed arrows). The loss function commonly used in style transfer consists of three parts: (i) *content loss* makes the synthesized image and the content image close in content features; (ii) *style loss* makes the synthesized image and style image close in style features; and (iii) *total variation loss* helps to reduce the noise in the synthesized image. Finally, when the model training is over, we output the model parameters of the style transfer to generate the final synthesized image.

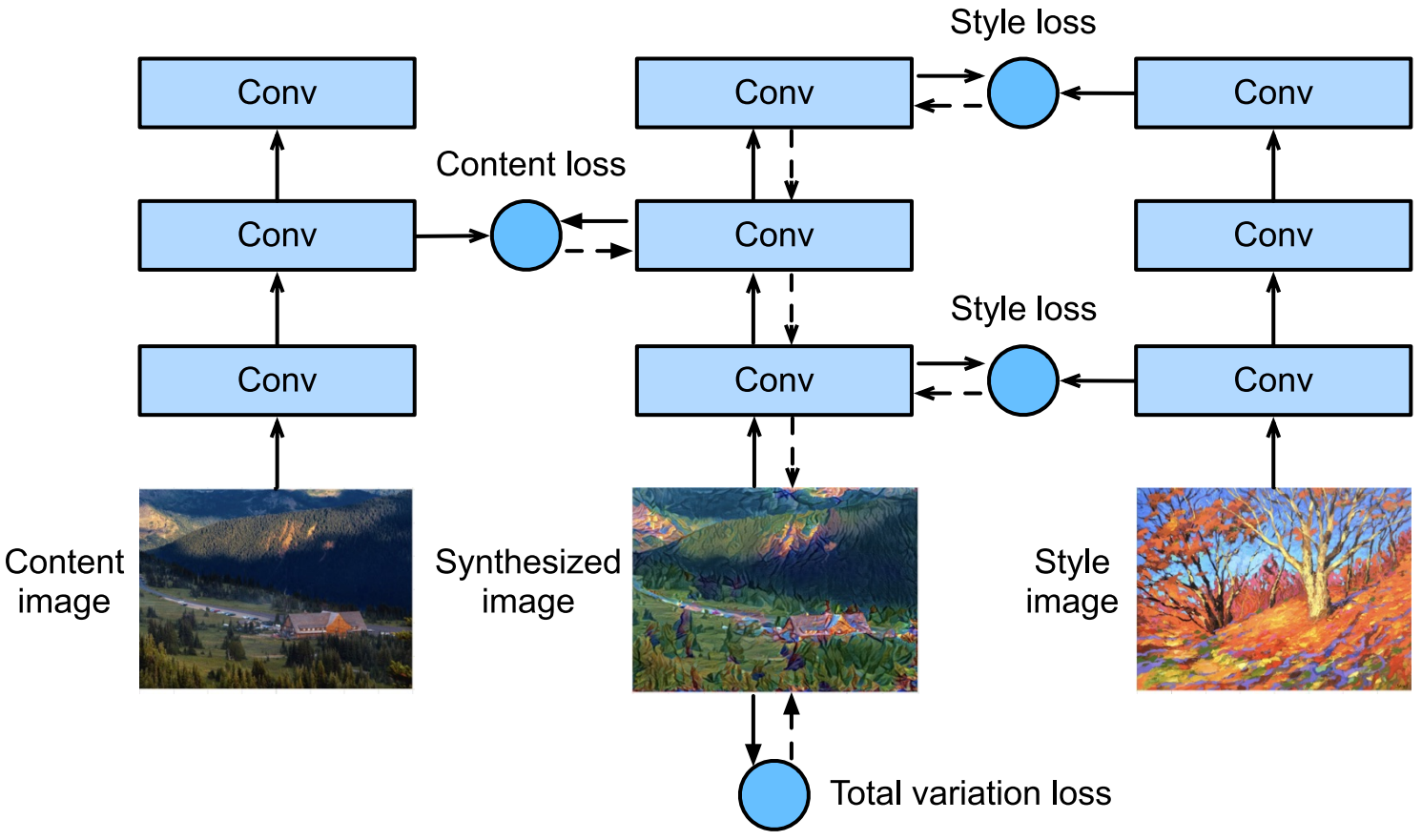

Fig. 14.12.2 CNN-based style transfer process. Solid lines show the direction of forward propagation and dotted lines show backward propagation.

In the following, we will explain the technical details of style transfer via a concrete experiment.

## 14.12.2 Reading the Content and Style Images

First, we read the content and style images. From their printed coordinate axes, we can tell that these images have different sizes.

```
%matplotlib inline
import torch
import torchvision
from torch import nn
from d2l import torch as d2l

d2l.set_figsize()
content_img = d2l.Image.open('../img/rainier.jpg')
d2l.plt.imshow(content_img);
```

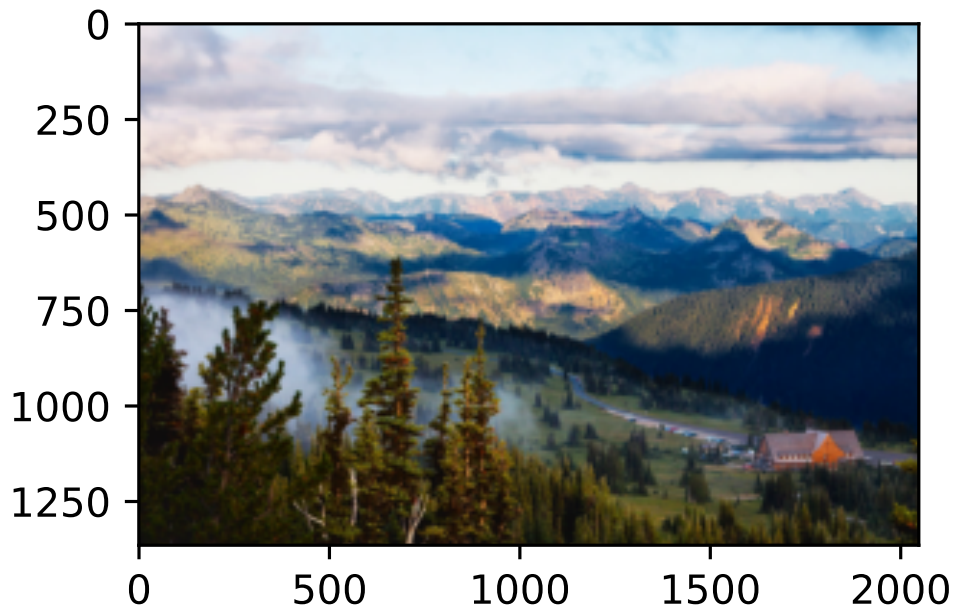

```
style_img = d2l.Image.open('../img/autumn-oak.jpg')
d2l.plt.imshow(style_img);
```

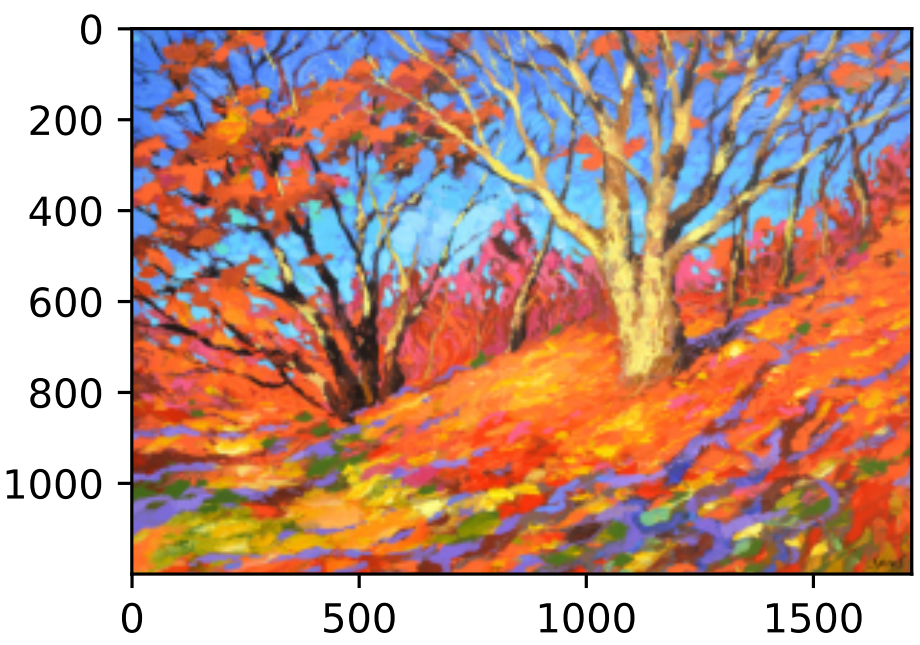

### 14.12.3 Preprocessing and Postprocessing

Below, we define two functions for preprocessing and postprocessing images. The `preprocess` function standardizes each of the three RGB channels of the input image and transforms the results into the CNN input format. The `postprocess` function restores the pixel values in the output image to their original values before standardization. Since the image printing function requires that each pixel has a floating point value from 0 to 1, we replace any value smaller than 0 or greater than 1 with 0 or 1, respectively.

```
rgb_mean = torch.tensor([0.485, 0.456, 0.406])
rgb_std = torch.tensor([0.229, 0.224, 0.225])

def preprocess(img, image_shape):
    transforms = torchvision.transforms.Compose([
        torchvision.transforms.Resize(image_shape),
        torchvision.transforms.ToTensor(),
        torchvision.transforms.Normalize(mean=rgb_mean, std=rgb_std)])
    return transforms(img).unsqueeze(0)

def postprocess(img):
    img = img[0].to(rgb_std.device)
    img = torch.clamp(img.permute(1, 2, 0) * rgb_std + rgb_mean, 0, 1)
    return torchvision.transforms.ToPILImage()(img.permute(2, 0, 1))
```

### 14.12.4 Extracting Features

We use the VGG-19 model pretrained on the ImageNet dataset to extract image features (Gatys *et al.*, 2016).

```
pretrained_net = torchvision.models.vgg19(pretrained=True)
```

In order to extract the content features and style features of the image, we can select the output of certain layers in the VGG network. Generally speaking, the closer to the input layer, the easier to extract details of the image, and vice versa, the easier to extract the global information of the image. In order to avoid excessively retaining the details of the content image in the synthesized image, we choose a VGG layer that is closer to the output as the *content layer* to output the content features of the image. We also select the output of different VGG layers for extracting local and global style features. These layers are also called *style layers*. As mentioned in Section 8.2, the VGG network uses 5 convolutional blocks. In the experiment, we choose the last convolutional layer of the fourth convolutional block as the content layer, and the first convolutional layer of each convolutional block as the style layer. The indices of these layers can be obtained by printing the `pretrained_net` instance.

```
style_layers, content_layers = [0, 5, 10, 19, 28], [25]
```

When extracting features using VGG layers, we only need to use all those from the input layer to the content layer or style layer that is closest to the output layer. Let's construct a new network instance `net`, which only retains all the VGG layers to be used for feature extraction.

```
net = nn.Sequential(*[pretrained_net.features[i] for i in
                      range(max(content_layers + style_layers) + 1)])
```

Given the input `X`, if we simply invoke the forward propagation `net(X)`, we can only get the output of the last layer. Since we also need the outputs of intermediate layers, we need to perform layer-by-layer computation and keep the content and style layer outputs.

```
def extract_features(X, content_layers, style_layers):
    contents = []
    styles = []
    for i in range(len(net)):
        X = net[i](X)
        if i in style_layers:
            styles.append(X)
        if i in content_layers:
            contents.append(X)
    return contents, styles
```

Two functions are defined below: the `get_contents` function extracts content features from the content image, and the `get_styles` function extracts style features from the style image. Since there is no need to update the model parameters of the pretrained VGG during training, we can extract the content and the style features even before the training starts. Since the synthesized image is a set of model parameters to be updated for style transfer, we can only extract the content and style features of the synthesized image by calling the `extract_features` function during training.

```
def get_contents(image_shape, device):
    content_X = preprocess(content_img, image_shape).to(device)
```

```
    contents_Y, _ = extract_features(content_X, content_layers, style_layers)
    return content_X, contents_Y

def get_styles(image_shape, device):
    style_X = preprocess(style_img, image_shape).to(device)
    _, styles_Y = extract_features(style_X, content_layers, style_layers)
    return style_X, styles_Y
```

## 14.12.5 Defining the Loss Function

Now we will describe the loss function for style transfer. The loss function consists of the content loss, style loss, and total variation loss.

### Content Loss

Similar to the loss function in linear regression, the content loss measures the difference in content features between the synthesized image and the content image via the squared loss function. The two inputs of the squared loss function are both outputs of the content layer computed by the `extract_features` function.

```
def content_loss(Y_hat, Y):
    # We detach the target content from the tree used to dynamically compute
    # the gradient: this is a stated value, not a variable. Otherwise the loss
    # will throw an error.
    return torch.square(Y_hat - Y.detach()).mean()
```

### Style Loss

Style loss, similar to content loss, also uses the squared loss function to measure the difference in style between the synthesized image and the style image. To express the style output of any style layer, we first use the `extract_features` function to compute the style layer output. Suppose that the output has 1 example, $c$ channels, height $h$, and width $w$, we can transform this output into matrix $\mathbf{X}$ with $c$ rows and $hw$ columns. This matrix can be thought of as the concatenation of $c$ vectors $\mathbf{x}_1, \ldots, \mathbf{x}_c$, each of which has a length of $hw$. Here, vector $\mathbf{x}_i$ represents the style feature of channel $i$.

In the *Gram matrix* of these vectors $\mathbf{X}\mathbf{X}^\top \in \mathbb{R}^{c \times c}$, element $x_{ij}$ in row $i$ and column $j$ is the dot product of vectors $\mathbf{x}_i$ and $\mathbf{x}_j$. It represents the correlation of the style features of channels $i$ and $j$. We use this Gram matrix to represent the style output of any style layer. Note that when the value of $hw$ is larger, it likely leads to larger values in the Gram matrix. Note also that the height and width of the Gram matrix are both the number of channels $c$. To allow style loss not to be affected by these values, the `gram` function below divides the Gram matrix by the number of its elements, i.e., $chw$.

```
def gram(X):
    num_channels, n = X.shape[1], X.numel() // X.shape[1]
    X = X.reshape((num_channels, n))
    return torch.matmul(X, X.T) / (num_channels * n)
```

Obviously, the two Gram matrix inputs of the squared loss function for style loss are based on the style layer outputs for the synthesized image and the style image. It is assumed here that the Gram matrix `gram_Y` based on the style image has been precomputed.

```
def style_loss(Y_hat, gram_Y):
    return torch.square(gram(Y_hat) - gram_Y.detach()).mean()
```

### Total Variation Loss

Sometimes, the learned synthesized image has a lot of high-frequency noise, i.e., particularly bright or dark pixels. One common noise reduction method is *total variation denoising*. Denote by $x_{i,j}$ the pixel value at coordinate $(i, j)$. Reducing total variation loss

$$\sum_{i,j} \left|x_{i,j} - x_{i+1,j}\right| + \left|x_{i,j} - x_{i,j+1}\right| \tag{14.12.1}$$

makes values of neighboring pixels on the synthesized image closer.

```
def tv_loss(Y_hat):
    return 0.5 * (torch.abs(Y_hat[:, :, 1:, :] - Y_hat[:, :, :-1, :]).mean() +
                  torch.abs(Y_hat[:, :, :, 1:] - Y_hat[:, :, :, :-1]).mean())
```

### Loss Function

The loss function of style transfer is the weighted sum of content loss, style loss, and total variation loss. By adjusting these weight hyperparameters, we can balance among content retention, style transfer, and noise reduction on the synthesized image.

```
content_weight, style_weight, tv_weight = 1, 1e4, 10

def compute_loss(X, contents_Y_hat, styles_Y_hat, contents_Y, styles_Y_gram):
    # Calculate the content, style, and total variance losses respectively
    contents_l = [content_loss(Y_hat, Y) * content_weight for Y_hat, Y in zip(
        contents_Y_hat, contents_Y)]
    styles_l = [style_loss(Y_hat, Y) * style_weight for Y_hat, Y in zip(
        styles_Y_hat, styles_Y_gram)]
    tv_l = tv_loss(X) * tv_weight
    # Add up all the losses
    l = sum(styles_l + contents_l + [tv_l])
    return contents_l, styles_l, tv_l, l
```

### 14.12.6 Initializing the Synthesized Image

In style transfer, the synthesized image is the only variable that needs to be updated during training. Thus, we can define a simple model, `SynthesizedImage`, and treat the synthesized image as the model parameters. In this model, forward propagation just returns the model parameters.

```
class SynthesizedImage(nn.Module):
    def __init__(self, img_shape, **kwargs):
        super(SynthesizedImage, self).__init__(**kwargs)
        self.weight = nn.Parameter(torch.rand(*img_shape))

    def forward(self):
        return self.weight
```

Next, we define the `get_inits` function. This function creates a synthesized image model instance and initializes it to the image `X`. Gram matrices for the style image at various style layers, `styles_Y_gram`, are computed prior to training.

```
def get_inits(X, device, lr, styles_Y):
    gen_img = SynthesizedImage(X.shape).to(device)
    gen_img.weight.data.copy_(X.data)
    trainer = torch.optim.Adam(gen_img.parameters(), lr=lr)
    styles_Y_gram = [gram(Y) for Y in styles_Y]
    return gen_img(), styles_Y_gram, trainer
```

### 14.12.7 Training

When training the model for style transfer, we continuously extract content features and style features of the synthesized image, and calculate the loss function. Below defines the training loop.

```
def train(X, contents_Y, styles_Y, device, lr, num_epochs, lr_decay_epoch):
    X, styles_Y_gram, trainer = get_inits(X, device, lr, styles_Y)
    scheduler = torch.optim.lr_scheduler.StepLR(trainer, lr_decay_epoch, 0.8)
    animator = d2l.Animator(xlabel='epoch', ylabel='loss',
                            xlim=[10, num_epochs],
                            legend=['content', 'style', 'TV'],
                            ncols=2, figsize=(7, 2.5))
    for epoch in range(num_epochs):
        trainer.zero_grad()
        contents_Y_hat, styles_Y_hat = extract_features(
            X, content_layers, style_layers)
        contents_l, styles_l, tv_l, l = compute_loss(
            X, contents_Y_hat, styles_Y_hat, contents_Y, styles_Y_gram)
        l.backward()
        trainer.step()
        scheduler.step()
        if (epoch + 1) % 10 == 0:
```

(continued from previous page)

```
        animator.axes[1].imshow(postprocess(X))
        animator.add(epoch + 1, [float(sum(contents_l)),
                                 float(sum(styles_l)), float(tv_l)])
    return X
```

Now we start to train the model. We rescale the height and width of the content and style images to 300 by 450 pixels. We use the content image to initialize the synthesized image.

```
device, image_shape = d2l.try_gpu(), (300, 450)  # PIL Image (h, w)
net = net.to(device)
content_X, contents_Y = get_contents(image_shape, device)
_, styles_Y = get_styles(image_shape, device)
output = train(content_X, contents_Y, styles_Y, device, 0.3, 500, 50)
```

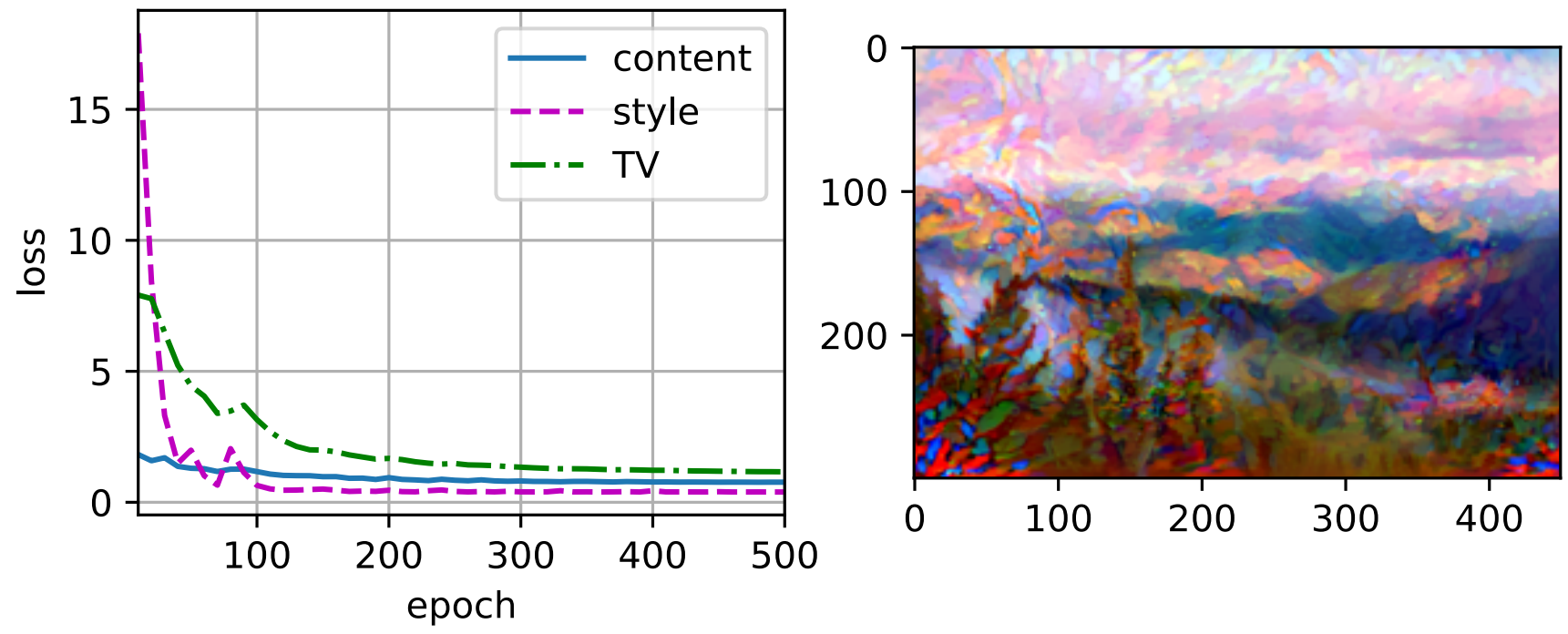

We can see that the synthesized image retains the scenery and objects of the content image, and transfers the color of the style image at the same time. For example, the synthesized image has blocks of color like those in the style image. Some of these blocks even have the subtle texture of brush strokes.

## 14.12.8 Summary

- The loss function commonly used in style transfer consists of three parts: (i) content loss makes the synthesized image and the content image close in content features; (ii) style loss makes the synthesized image and style image close in style features; and (iii) total variation loss helps to reduce the noise in the synthesized image.
- We can use a pretrained CNN to extract image features and minimize the loss function to continuously update the synthesized image as model parameters during training.
- We use Gram matrices to represent the style outputs from the style layers.

## 14.12.9 Exercises

1. How does the output change when you select different content and style layers?
2. Adjust the weight hyperparameters in the loss function. Does the output retain more content or have less noise?
3. Use different content and style images. Can you create more interesting synthesized images?
4. Can we apply style transfer for text? Hint: you may refer to the survey paper by Hu *et al.* (2022).

Discussions[224].

## 14.13 Image Classification (CIFAR-10) on Kaggle

So far, we have been using high-level APIs of deep learning frameworks to directly obtain image datasets in tensor format. However, custom image datasets often come in the form of image files. In this section, we will start from raw image files, and organize, read, then transform them into tensor format step by step.

We experimented with the CIFAR-10 dataset in Section 14.1, which is an important dataset in computer vision. In this section, we will apply the knowledge we learned in previous sections to practice the Kaggle competition of CIFAR-10 image classification. The web address of the competition is https://www.kaggle.com/c/cifar-10

Fig. 14.13.1 shows the information on the competition's webpage. In order to submit the results, you need to register a Kaggle account.

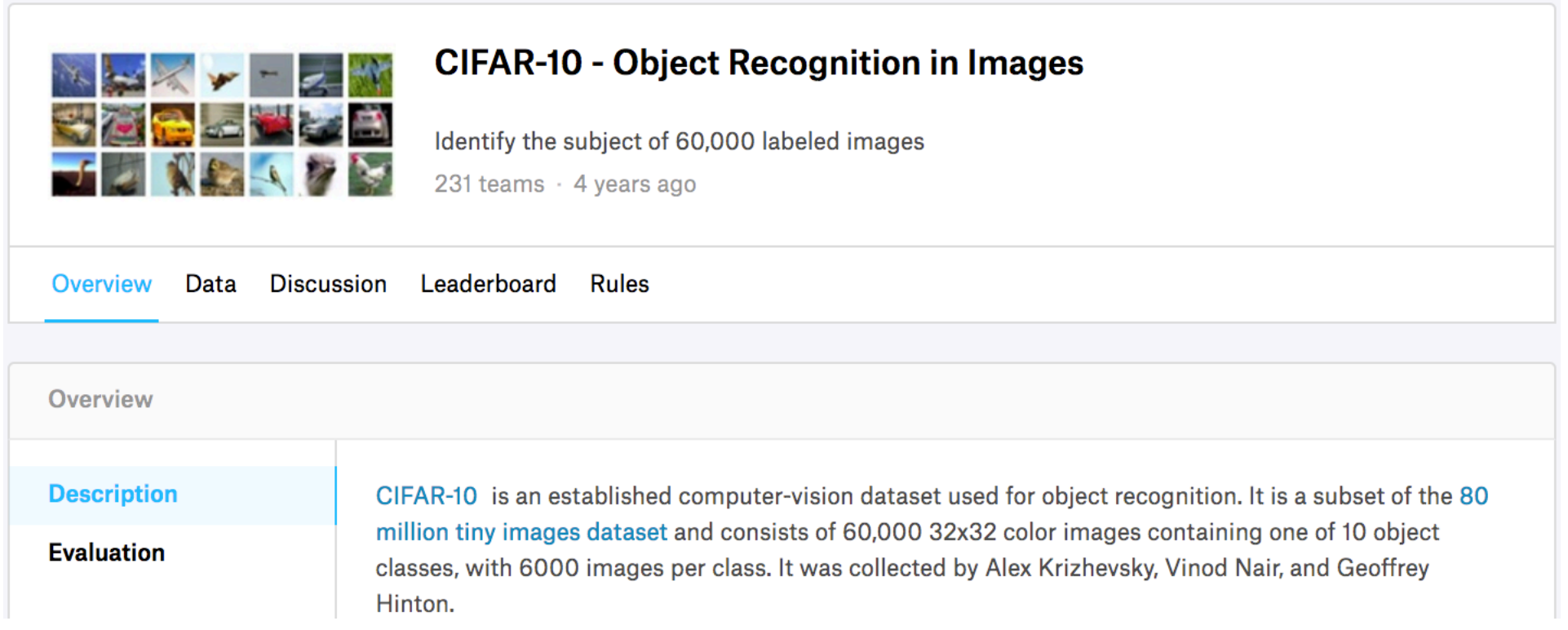

Fig. 14.13.1 CIFAR-10 image classification competition webpage information. The competition dataset can be obtained by clicking the "Data" tab.

```
import collections
import math
import os
import shutil
import pandas as pd
import torch
import torchvision
from torch import nn
from d2l import torch as d2l
```

## 14.13.1 Obtaining and Organizing the Dataset

The competition dataset is divided into a training set and a test set, which contain 50000 and 300000 images, respectively. In the test set, 10000 images will be used for evaluation, while the remaining 290000 images will not be evaluated: they are included just to make it hard to cheat with *manually* labeled results of the test set. The images in this dataset are all png color (RGB channels) image files, whose height and width are both 32 pixels. The images cover a total of 10 categories, namely airplanes, cars, birds, cats, deer, dogs, frogs, horses, boats, and trucks. The upper-left corner of Fig. 14.13.1 shows some images of airplanes, cars, and birds in the dataset.

### Downloading the Dataset

After logging in to Kaggle, we can click the "Data" tab on the CIFAR-10 image classification competition webpage shown in Fig. 14.13.1 and download the dataset by clicking the "Download All" button. After unzipping the downloaded file in `../data`, and unzipping `train.7z` and `test.7z` inside it, you will find the entire dataset in the following paths:

- `../data/cifar-10/train/[1-50000].png`
- `../data/cifar-10/test/[1-300000].png`
- `../data/cifar-10/trainLabels.csv`
- `../data/cifar-10/sampleSubmission.csv`

where the `train` and `test` directories contain the training and testing images, respectively, `trainLabels.csv` provides labels for the training images, and `sample_submission.csv` is a sample submission file.

To make it easier to get started, we provide a small-scale sample of the dataset that contains the first 1000 training images and 5 random testing images. To use the full dataset of the Kaggle competition, you need to set the following demo variable to `False`.

```
#@save
d2l.DATA_HUB['cifar10_tiny'] = (d2l.DATA_URL + 'kaggle_cifar10_tiny.zip',
```

```
                                '2068874e4b9a9f0fb07ebe0ad2b29754449ccacd')

# If you use the full dataset downloaded for the Kaggle competition, set
# `demo` to False
demo = True

if demo:
    data_dir = d2l.download_extract('cifar10_tiny')
else:
    data_dir = '../data/cifar-10/'
```

```
Downloading ../data/kaggle_cifar10_tiny.zip from http://d2l-data.s3-accelerate.
↪amazonaws.com/kaggle_cifar10_tiny.zip...
```

### Organizing the Dataset

We need to organize datasets to facilitate model training and testing. Let's first read the labels from the csv file. The following function returns a dictionary that maps the non-extension part of the filename to its label.

```
#@save
def read_csv_labels(fname):
    """Read `fname` to return a filename to label dictionary."""
    with open(fname, 'r') as f:
        # Skip the file header line (column name)
        lines = f.readlines()[1:]
    tokens = [l.rstrip().split(',') for l in lines]
    return dict(((name, label) for name, label in tokens))

labels = read_csv_labels(os.path.join(data_dir, 'trainLabels.csv'))
print('# training examples:', len(labels))
print('# classes:', len(set(labels.values())))
```

```
# training examples: 1000
# classes: 10
```

Next, we define the `reorg_train_valid` function to split the validation set out of the original training set. The argument `valid_ratio` in this function is the ratio of the number of examples in the validation set to the number of examples in the original training set. More concretely, let $n$ be the number of images of the class with the least examples, and $r$ be the ratio. The validation set will split out $\max(\lfloor nr \rfloor, 1)$ images for each class. Let's use `valid_ratio=0.1` as an example. Since the original training set has 50000 images, there will be 45000 images used for training in the path `train_valid_test/train`, while the other 5000 images will be split out as validation set in the path `train_valid_test/valid`. After organizing the dataset, images of the same class will be placed under the same folder.

```
#@save
def copyfile(filename, target_dir):
    """Copy a file into a target directory."""
    os.makedirs(target_dir, exist_ok=True)
    shutil.copy(filename, target_dir)

#@save
def reorg_train_valid(data_dir, labels, valid_ratio):
    """Split the validation set out of the original training set."""
    # The number of examples of the class that has the fewest examples in the
    # training dataset
    n = collections.Counter(labels.values()).most_common()[-1][1]
    # The number of examples per class for the validation set
    n_valid_per_label = max(1, math.floor(n * valid_ratio))
    label_count = {}
    for train_file in os.listdir(os.path.join(data_dir, 'train')):
        label = labels[train_file.split('.')[0]]
        fname = os.path.join(data_dir, 'train', train_file)
        copyfile(fname, os.path.join(data_dir, 'train_valid_test',
                                     'train_valid', label))
        if label not in label_count or label_count[label] < n_valid_per_label:
            copyfile(fname, os.path.join(data_dir, 'train_valid_test',
                                         'valid', label))
            label_count[label] = label_count.get(label, 0) + 1
        else:
            copyfile(fname, os.path.join(data_dir, 'train_valid_test',
                                         'train', label))
    return n_valid_per_label
```

The `reorg_test` function below organizes the testing set for data loading during prediction.

```
#@save
def reorg_test(data_dir):
    """Organize the testing set for data loading during prediction."""
    for test_file in os.listdir(os.path.join(data_dir, 'test')):
        copyfile(os.path.join(data_dir, 'test', test_file),
                 os.path.join(data_dir, 'train_valid_test', 'test',
                              'unknown'))
```

Finally, we use a function to invoke the `read_csv_labels`, `reorg_train_valid`, and `reorg_test` functions defined above.

```
def reorg_cifar10_data(data_dir, valid_ratio):
    labels = read_csv_labels(os.path.join(data_dir, 'trainLabels.csv'))
    reorg_train_valid(data_dir, labels, valid_ratio)
    reorg_test(data_dir)
```

Here we only set the batch size to 32 for the small-scale sample of the dataset. When training and testing the complete dataset of the Kaggle competition, `batch_size` should be set to a larger integer, such as 128. We split out 10% of the training examples as the validation set for tuning hyperparameters.

```
batch_size = 32 if demo else 128
valid_ratio = 0.1
reorg_cifar10_data(data_dir, valid_ratio)
```

## 14.13.2 Image Augmentation

We use image augmentation to address overfitting. For example, images can be flipped horizontally at random during training. We can also perform standardization for the three RGB channels of color images. Below lists some of these operations that you can tweak.

```
transform_train = torchvision.transforms.Compose([
    # Scale the image up to a square of 40 pixels in both height and width
    torchvision.transforms.Resize(40),
    # Randomly crop a square image of 40 pixels in both height and width to
    # produce a small square of 0.64 to 1 times the area of the original
    # image, and then scale it to a square of 32 pixels in both height and
    # width
    torchvision.transforms.RandomResizedCrop(32, scale=(0.64, 1.0),
                                              ratio=(1.0, 1.0)),
    torchvision.transforms.RandomHorizontalFlip(),
    torchvision.transforms.ToTensor(),
    # Standardize each channel of the image
    torchvision.transforms.Normalize([0.4914, 0.4822, 0.4465],
                                     [0.2023, 0.1994, 0.2010])])
```

During testing, we only perform standardization on images so as to remove randomness in the evaluation results.

```
transform_test = torchvision.transforms.Compose([
    torchvision.transforms.ToTensor(),
    torchvision.transforms.Normalize([0.4914, 0.4822, 0.4465],
                                     [0.2023, 0.1994, 0.2010])])
```

## 14.13.3 Reading the Dataset

Next, we read the organized dataset consisting of raw image files. Each example includes an image and a label.

```
train_ds, train_valid_ds = [torchvision.datasets.ImageFolder(
    os.path.join(data_dir, 'train_valid_test', folder),
    transform=transform_train) for folder in ['train', 'train_valid']]

valid_ds, test_ds = [torchvision.datasets.ImageFolder(
    os.path.join(data_dir, 'train_valid_test', folder),
    transform=transform_test) for folder in ['valid', 'test']]
```

During training, we need to specify all the image augmentation operations defined above.

When the validation set is used for model evaluation during hyperparameter tuning, no randomness from image augmentation should be introduced. Before final prediction, we train the model on the combined training set and validation set to make full use of all the labeled data.

```
train_iter, train_valid_iter = [torch.utils.data.DataLoader(
    dataset, batch_size, shuffle=True, drop_last=True)
    for dataset in (train_ds, train_valid_ds)]

valid_iter = torch.utils.data.DataLoader(valid_ds, batch_size, shuffle=False,
                                         drop_last=True)

test_iter = torch.utils.data.DataLoader(test_ds, batch_size, shuffle=False,
                                        drop_last=False)
```

## 14.13.4 Defining the Model

We define the ResNet-18 model described in Section 8.6.

```
def get_net():
    num_classes = 10
    net = d2l.resnet18(num_classes, 3)
    return net

loss = nn.CrossEntropyLoss(reduction="none")
```

## 14.13.5 Defining the Training Function

We will select models and tune hyperparameters according to the model's performance on the validation set. In the following, we define the model training function `train`.

```
def train(net, train_iter, valid_iter, num_epochs, lr, wd, devices, lr_period,
          lr_decay):
    trainer = torch.optim.SGD(net.parameters(), lr=lr, momentum=0.9,
                              weight_decay=wd)
    scheduler = torch.optim.lr_scheduler.StepLR(trainer, lr_period, lr_decay)
    num_batches, timer = len(train_iter), d2l.Timer()
    legend = ['train loss', 'train acc']
    if valid_iter is not None:
        legend.append('valid acc')
    animator = d2l.Animator(xlabel='epoch', xlim=[1, num_epochs],
                            legend=legend)
    net = nn.DataParallel(net, device_ids=devices).to(devices[0])
    for epoch in range(num_epochs):
        net.train()
        metric = d2l.Accumulator(3)
        for i, (features, labels) in enumerate(train_iter):
            timer.start()
            l, acc = d2l.train_batch_ch13(net, features, labels,
```

```
                                       loss, trainer, devices)
            metric.add(l, acc, labels.shape[0])
            timer.stop()
            if (i + 1) % (num_batches // 5) == 0 or i == num_batches - 1:
                animator.add(epoch + (i + 1) / num_batches,
                             (metric[0] / metric[2], metric[1] / metric[2],
                              None))
        if valid_iter is not None:
            valid_acc = d2l.evaluate_accuracy_gpu(net, valid_iter)
            animator.add(epoch + 1, (None, None, valid_acc))
        scheduler.step()
    measures = (f'train loss {metric[0] / metric[2]:.3f}, '
                f'train acc {metric[1] / metric[2]:.3f}')
    if valid_iter is not None:
        measures += f', valid acc {valid_acc:.3f}'
    print(measures + f'\n{metric[2] * num_epochs / timer.sum():.1f}'
          f' examples/sec on {str(devices)}')
```

## 14.13.6 Training and Validating the Model

Now, we can train and validate the model. All the following hyperparameters can be tuned. For example, we can increase the number of epochs. When `lr_period` and `lr_decay` are set to 4 and 0.9, respectively, the learning rate of the optimization algorithm will be multiplied by 0.9 after every 4 epochs. Just for ease of demonstration, we only train 20 epochs here.

```
devices, num_epochs, lr, wd = d2l.try_all_gpus(), 20, 2e-4, 5e-4
lr_period, lr_decay, net = 4, 0.9, get_net()
net(next(iter(train_iter))[0])
train(net, train_iter, valid_iter, num_epochs, lr, wd, devices, lr_period,
      lr_decay)
```

```
train loss 0.727, train acc 0.738, valid acc 0.484
613.0 examples/sec on [device(type='cuda', index=0), device(type='cuda',␣
↪index=1)]
```

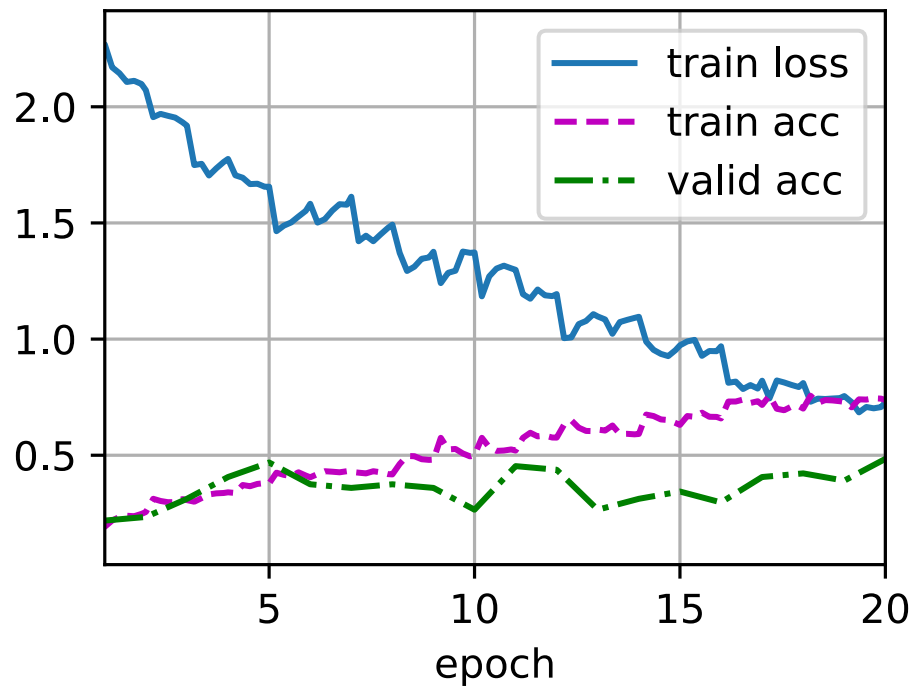

### 14.13.7 Classifying the Testing Set and Submitting Results on Kaggle

After obtaining a promising model with hyperparameters, we use all the labeled data (including the validation set) to retrain the model and classify the testing set.

```
net, preds = get_net(), []
net(next(iter(train_valid_iter))[0])
train(net, train_valid_iter, None, num_epochs, lr, wd, devices, lr_period,
      lr_decay)

for X, _ in test_iter:
    y_hat = net(X.to(devices[0]))
    preds.extend(y_hat.argmax(dim=1).type(torch.int32).cpu().numpy())
sorted_ids = list(range(1, len(test_ds) + 1))
sorted_ids.sort(key=lambda x: str(x))
df = pd.DataFrame({'id': sorted_ids, 'label': preds})
df['label'] = df['label'].apply(lambda x: train_valid_ds.classes[x])
df.to_csv('submission.csv', index=False)
```

```
train loss 0.792, train acc 0.725
806.6 examples/sec on [device(type='cuda', index=0), device(type='cuda',␣
↪index=1)]
```

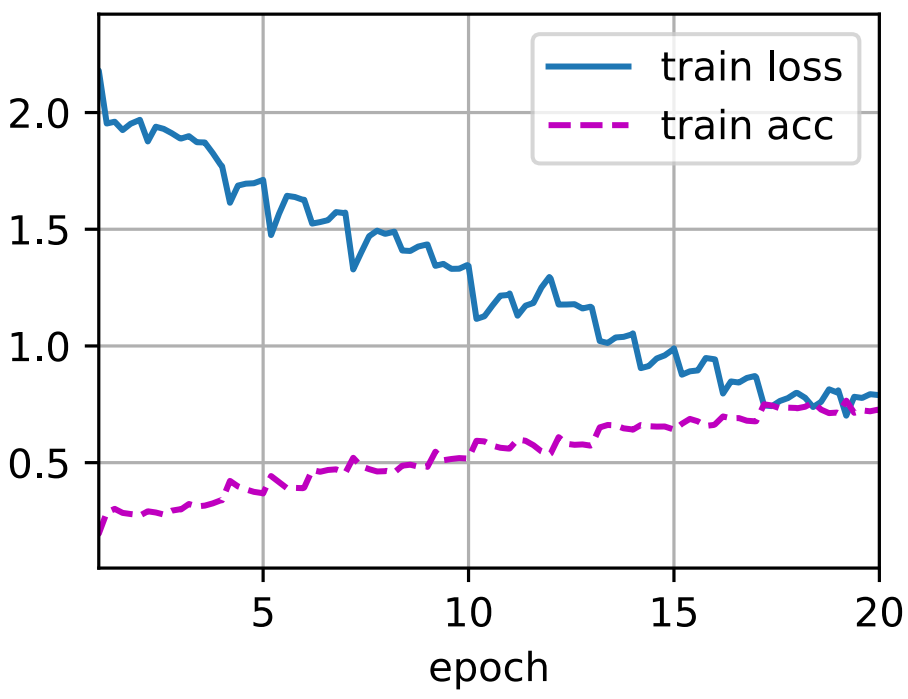

The above code will generate a `submission.csv` file, whose format meets the requirement of the Kaggle competition. The method for submitting results to Kaggle is similar to that in Section 5.7.

### 14.13.8 Summary

- We can read datasets containing raw image files after organizing them into the required format.
- We can use convolutional neural networks and image augmentation in an image classification competition.

### 14.13.9 Exercises

1. Use the complete CIFAR-10 dataset for this Kaggle competition. Set hyperparameters as `batch_size = 128`, `num_epochs = 100`, `lr = 0.1`, `lr_period = 50`, and `lr_decay = 0.1`. See what accuracy and ranking you can achieve in this competition. Can you further improve them?

2. What accuracy can you get when not using image augmentation?

Discussions[225].

## 14.14 Dog Breed Identification (ImageNet Dogs) on Kaggle

In this section, we will practice the dog breed identification problem on Kaggle. The web address of this competition is https://www.kaggle.com/c/dog-breed-identification

In this competition, 120 different breeds of dogs will be recognized. In fact, the dataset for this competition is a subset of the ImageNet dataset. Unlike the images in the CIFAR-10 dataset in Section 14.13, the images in the ImageNet dataset are both higher and wider in varying dimensions. Fig. 14.14.1 shows the information on the competition's webpage. You need a Kaggle account to submit your results.

```
import os
import torch
import torchvision
from torch import nn
from d2l import torch as d2l
```

### 14.14.1 Obtaining and Organizing the Dataset

The competition dataset is divided into a training set and a test set, which contain 10222 and 10357 JPEG images of three RGB (color) channels, respectively. Among the training dataset, there are 120 breeds of dogs such as Labradors, Poodles, Dachshunds, Samoyeds, Huskies, Chihuahuas, and Yorkshire Terriers.

#### Downloading the Dataset

After logging into Kaggle, you can click on the "Data" tab on the competition webpage shown in Fig. 14.14.1 and download the dataset by clicking the "Download All" button. After un-

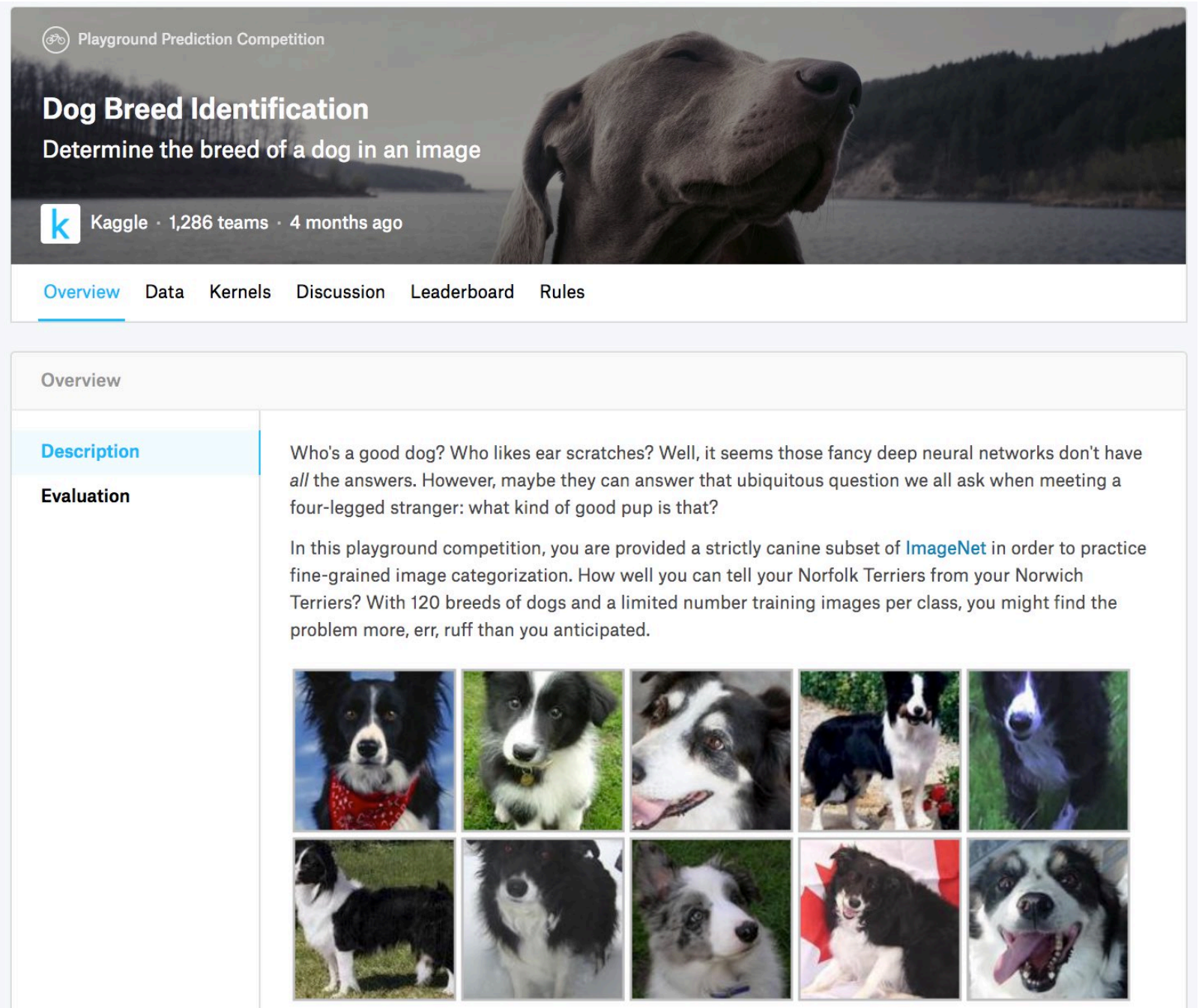

Fig. 14.14.1 The dog breed identification competition website. The competition dataset can be obtained by clicking the "Data" tab.

zipping the downloaded file in `../data`, you will find the entire dataset in the following paths:

- ../data/dog-breed-identification/labels.csv
- ../data/dog-breed-identification/sample_submission.csv
- ../data/dog-breed-identification/train
- ../data/dog-breed-identification/test

You may have noticed that the above structure is similar to that of the CIFAR-10 competition in Section 14.13, where folders `train/` and `test/` contain training and testing dog images, respectively, and `labels.csv` contains the labels for the training images. Similarly, to make it easier to get started, we provide a small sample of the dataset mentioned above: `train_valid_test_tiny.zip`. If you are going to use the full dataset for the Kaggle competition, you need to change the `demo` variable below to `False`.

```
#@save
d2l.DATA_HUB['dog_tiny'] = (d2l.DATA_URL + 'kaggle_dog_tiny.zip',
                            '0cb91d09b814ecdc07b50f31f8dcad3e81d6a86d')

# If you use the full dataset downloaded for the Kaggle competition, change
# the variable below to `False`
```

```
demo = True
if demo:
    data_dir = d2l.download_extract('dog_tiny')
else:
    data_dir = os.path.join('..', 'data', 'dog-breed-identification')
```

### Organizing the Dataset

We can organize the dataset similarly to what we did in Section 14.13, namely splitting out a validation set from the original training set, and moving images into subfolders grouped by labels.

The `reorg_dog_data` function below reads the training data labels, splits out the validation set, and organizes the training set.

```
def reorg_dog_data(data_dir, valid_ratio):
    labels = d2l.read_csv_labels(os.path.join(data_dir, 'labels.csv'))
    d2l.reorg_train_valid(data_dir, labels, valid_ratio)
    d2l.reorg_test(data_dir)

batch_size = 32 if demo else 128
valid_ratio = 0.1
reorg_dog_data(data_dir, valid_ratio)
```

## 14.14.2 Image Augmentation

Recall that this dog breed dataset is a subset of the ImageNet dataset, whose images are larger than those of the CIFAR-10 dataset in Section 14.13. The following lists a few image augmentation operations that might be useful for relatively larger images.

```
transform_train = torchvision.transforms.Compose([
    # Randomly crop the image to obtain an image with an area of 0.08 to 1 of
    # the original area and height-to-width ratio between 3/4 and 4/3. Then,
    # scale the image to create a new 224 x 224 image
    torchvision.transforms.RandomResizedCrop(224, scale=(0.08, 1.0),
                                             ratio=(3.0/4.0, 4.0/3.0)),
    torchvision.transforms.RandomHorizontalFlip(),
    # Randomly change the brightness, contrast, and saturation
    torchvision.transforms.ColorJitter(brightness=0.4,
                                       contrast=0.4,
                                       saturation=0.4),
    # Add random noise
    torchvision.transforms.ToTensor(),
    # Standardize each channel of the image
    torchvision.transforms.Normalize([0.485, 0.456, 0.406],
                                     [0.229, 0.224, 0.225])])
```

During prediction, we only use image preprocessing operations without randomness.

```
transform_test = torchvision.transforms.Compose([
    torchvision.transforms.Resize(256),
    # Crop a 224 x 224 square area from the center of the image
    torchvision.transforms.CenterCrop(224),
    torchvision.transforms.ToTensor(),
    torchvision.transforms.Normalize([0.485, 0.456, 0.406],
                                     [0.229, 0.224, 0.225])])
```

### 14.14.3 Reading the Dataset

As in Section 14.13, we can read the organized dataset consisting of raw image files.

```
train_ds, train_valid_ds = [torchvision.datasets.ImageFolder(
    os.path.join(data_dir, 'train_valid_test', folder),
    transform=transform_train) for folder in ['train', 'train_valid']]

valid_ds, test_ds = [torchvision.datasets.ImageFolder(
    os.path.join(data_dir, 'train_valid_test', folder),
    transform=transform_test) for folder in ['valid', 'test']]
```

Below we create data iterator instances the same way as in Section 14.13.

```
train_iter, train_valid_iter = [torch.utils.data.DataLoader(
    dataset, batch_size, shuffle=True, drop_last=True)
    for dataset in (train_ds, train_valid_ds)]

valid_iter = torch.utils.data.DataLoader(valid_ds, batch_size, shuffle=False,
                                         drop_last=True)

test_iter = torch.utils.data.DataLoader(test_ds, batch_size, shuffle=False,
                                        drop_last=False)
```

### 14.14.4 Fine-Tuning a Pretrained Model

Again, the dataset for this competition is a subset of the ImageNet dataset. Therefore, we can use the approach discussed in Section 14.2 to select a model pretrained on the full ImageNet dataset and use it to extract image features to be fed into a custom small-scale output network. High-level APIs of deep learning frameworks provide a wide range of models pretrained on the ImageNet dataset. Here, we choose a pretrained ResNet-34 model, where we simply reuse the input of this model's output layer (i.e., the extracted features). Then we can replace the original output layer with a small custom output network that can be trained, such as stacking two fully connected layers. Different from the experiment in Section 14.2, the following does not retrain the pretrained model used for feature extraction. This reduces training time and memory for storing gradients.

Recall that we standardized images using the means and standard deviations of the three RGB

channels for the full ImageNet dataset. In fact, this is also consistent with the standardization operation by the pretrained model on ImageNet.

```
def get_net(devices):
    finetune_net = nn.Sequential()
    finetune_net.features = torchvision.models.resnet34(pretrained=True)
    # Define a new output network (there are 120 output categories)
    finetune_net.output_new = nn.Sequential(nn.Linear(1000, 256),
                                            nn.ReLU(),
                                            nn.Linear(256, 120))
    # Move the model to devices
    finetune_net = finetune_net.to(devices[0])
    # Freeze parameters of feature layers
    for param in finetune_net.features.parameters():
        param.requires_grad = False
    return finetune_net
```

Before calculating the loss, we first obtain the input of the pretrained model's output layer, i.e., the extracted feature. Then we use this feature as input for our small custom output network to calculate the loss.

```
loss = nn.CrossEntropyLoss(reduction='none')

def evaluate_loss(data_iter, net, devices):
    l_sum, n = 0.0, 0
    for features, labels in data_iter:
        features, labels = features.to(devices[0]), labels.to(devices[0])
        outputs = net(features)
        l = loss(outputs, labels)
        l_sum += l.sum()
        n += labels.numel()
    return l_sum / n
```

## 14.14.5 Defining the Training Function

We will select the model and tune hyperparameters according to the model's performance on the validation set. The model training function `train` only iterates parameters of the small custom output network.

```
def train(net, train_iter, valid_iter, num_epochs, lr, wd, devices, lr_period,
          lr_decay):
    # Only train the small custom output network
    net = nn.DataParallel(net, device_ids=devices).to(devices[0])
    trainer = torch.optim.SGD((param for param in net.parameters()
                               if param.requires_grad), lr=lr,
                              momentum=0.9, weight_decay=wd)
    scheduler = torch.optim.lr_scheduler.StepLR(trainer, lr_period, lr_decay)
    num_batches, timer = len(train_iter), d2l.Timer()
    legend = ['train loss']
    if valid_iter is not None:
```

```
        legend.append('valid loss')
    animator = d2l.Animator(xlabel='epoch', xlim=[1, num_epochs],
                            legend=legend)
    for epoch in range(num_epochs):
        metric = d2l.Accumulator(2)
        for i, (features, labels) in enumerate(train_iter):
            timer.start()
            features, labels = features.to(devices[0]), labels.to(devices[0])
            trainer.zero_grad()
            output = net(features)
            l = loss(output, labels).sum()
            l.backward()
            trainer.step()
            metric.add(l, labels.shape[0])
            timer.stop()
            if (i + 1) % (num_batches // 5) == 0 or i == num_batches - 1:
                animator.add(epoch + (i + 1) / num_batches,
                             (metric[0] / metric[1], None))
        measures = f'train loss {metric[0] / metric[1]:.3f}'
        if valid_iter is not None:
            valid_loss = evaluate_loss(valid_iter, net, devices)
            animator.add(epoch + 1, (None, valid_loss.detach().cpu()))
        scheduler.step()
    if valid_iter is not None:
        measures += f', valid loss {valid_loss:.3f}'
    print(measures + f'\n{metric[1] * num_epochs / timer.sum():.1f}'
          f' examples/sec on {str(devices)}')
```

## 14.14.6 Training and Validating the Model

Now we can train and validate the model. The following hyperparameters are all tunable. For example, the number of epochs can be increased. Because `lr_period` and `lr_decay` are set to 2 and 0.9, respectively, the learning rate of the optimization algorithm will be multiplied by 0.9 after every 2 epochs.

```
devices, num_epochs, lr, wd = d2l.try_all_gpus(), 10, 1e-4, 1e-4
lr_period, lr_decay, net = 2, 0.9, get_net(devices)
train(net, train_iter, valid_iter, num_epochs, lr, wd, devices, lr_period,
      lr_decay)
```

```
train loss 1.230, valid loss 1.313
390.8 examples/sec on [device(type='cuda', index=0), device(type='cuda',␣
↪index=1)]
```

## 14.14.7 Classifying the Testing Set and Submitting Results on Kaggle

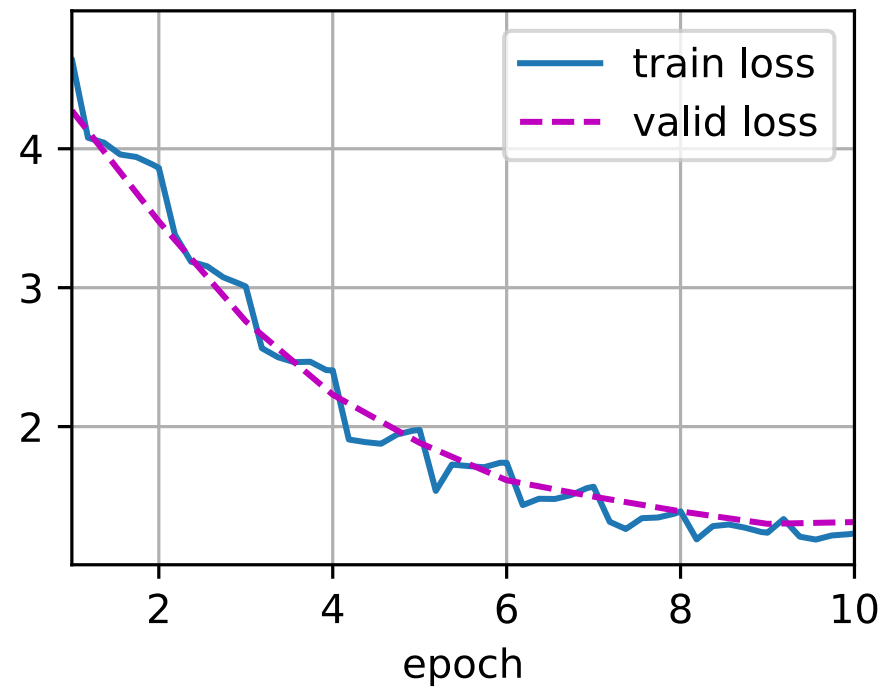

Similar to the final step in Section 14.13, in the end all the labeled data (including the validation set) are used for training the model and classifying the testing set. We will use the trained custom output network for classification.

```
net = get_net(devices)
train(net, train_valid_iter, None, num_epochs, lr, wd, devices, lr_period,
      lr_decay)

preds = []
for data, label in test_iter:
    output = torch.nn.functional.softmax(net(data.to(devices[0])), dim=1)
    preds.extend(output.cpu().detach().numpy())
ids = sorted(os.listdir(
    os.path.join(data_dir, 'train_valid_test', 'test', 'unknown')))
with open('submission.csv', 'w') as f:
    f.write('id,' + ','.join(train_valid_ds.classes) + '\n')
    for i, output in zip(ids, preds):
        f.write(i.split('.')[0] + ',' + ','.join(
            [str(num) for num in output]) + '\n')
```

```
train loss 1.151
582.9 examples/sec on [device(type='cuda', index=0), device(type='cuda',␣
↪index=1)]
```

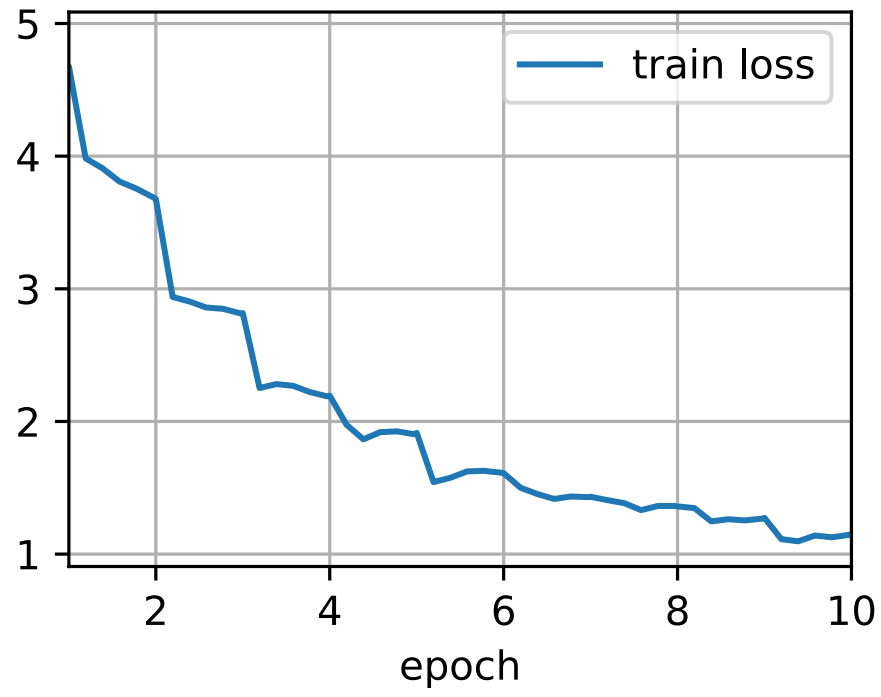

The above code will generate a `submission.csv` file to be submitted to Kaggle in the same way described in Section 5.7.

### 14.14.8 Summary

- Images in the ImageNet dataset are larger (with varying dimensions) than CIFAR-10 images. We may modify image augmentation operations for tasks on a different dataset.
- To classify a subset of the ImageNet dataset, we can leverage pre-trained models on the full ImageNet dataset to extract features and only train a custom small-scale output network. This will lead to less computational time and memory cost.

### 14.14.9 Exercises

1. When using the full Kaggle competition dataset, what results can you achieve when you increase `batch_size` (batch size) and `num_epochs` (number of epochs) while setting some other hyperparameters as `lr = 0.01`, `lr_period = 10`, and `lr_decay = 0.1`?
2. Do you get better results if you use a deeper pretrained model? How do you tune hyperparameters? Can you further improve the results?

Discussions[226].

# 15 Natural Language Processing: Pretraining

Humans need to communicate. Out of this basic need of the human condition, a vast amount of written text has been generated on an everyday basis. Given rich text in social media, chat apps, emails, product reviews, news articles, research papers, and books, it becomes vital to enable computers to understand them to offer assistance or make decisions based on human languages.

*Natural language processing* studies interactions between computers and humans using natural languages. In practice, it is very common to use natural language processing techniques to process and analyze text (human natural language) data, such as language models in Section 9.3 and machine translation models in Section 10.5.

To understand text, we can begin by learning its representations. Leveraging the existing text sequences from large corpora, *self-supervised learning* has been extensively used to pretrain text representations, such as by predicting some hidden part of the text using some other part of their surrounding text. In this way, models learn through supervision from *massive* text data without *expensive* labeling efforts!

As we will see in this chapter, when treating each word or subword as an individual token, the representation of each token can be pretrained using word2vec, GloVe, or subword embedding models on large corpora. After pretraining, representation of each token can be a vector, however, it remains the same no matter what the context is. For instance, the vector representation of "bank" is the same in both "go to the bank to deposit some money" and "go to the bank to sit down". Thus, many more recent pretraining models adapt representation of the same token to different contexts. Among them is BERT, a much deeper self-supervised model based on the Transformer encoder. In this chapter, we will focus on how to pretrain such representations for text, as highlighted in Fig. 15.1.

For sight of the big picture, Fig. 15.1 shows that the pretrained text representations can be fed to a variety of deep learning architectures for different downstream natural language processing applications. We will cover them in Chapter 16.

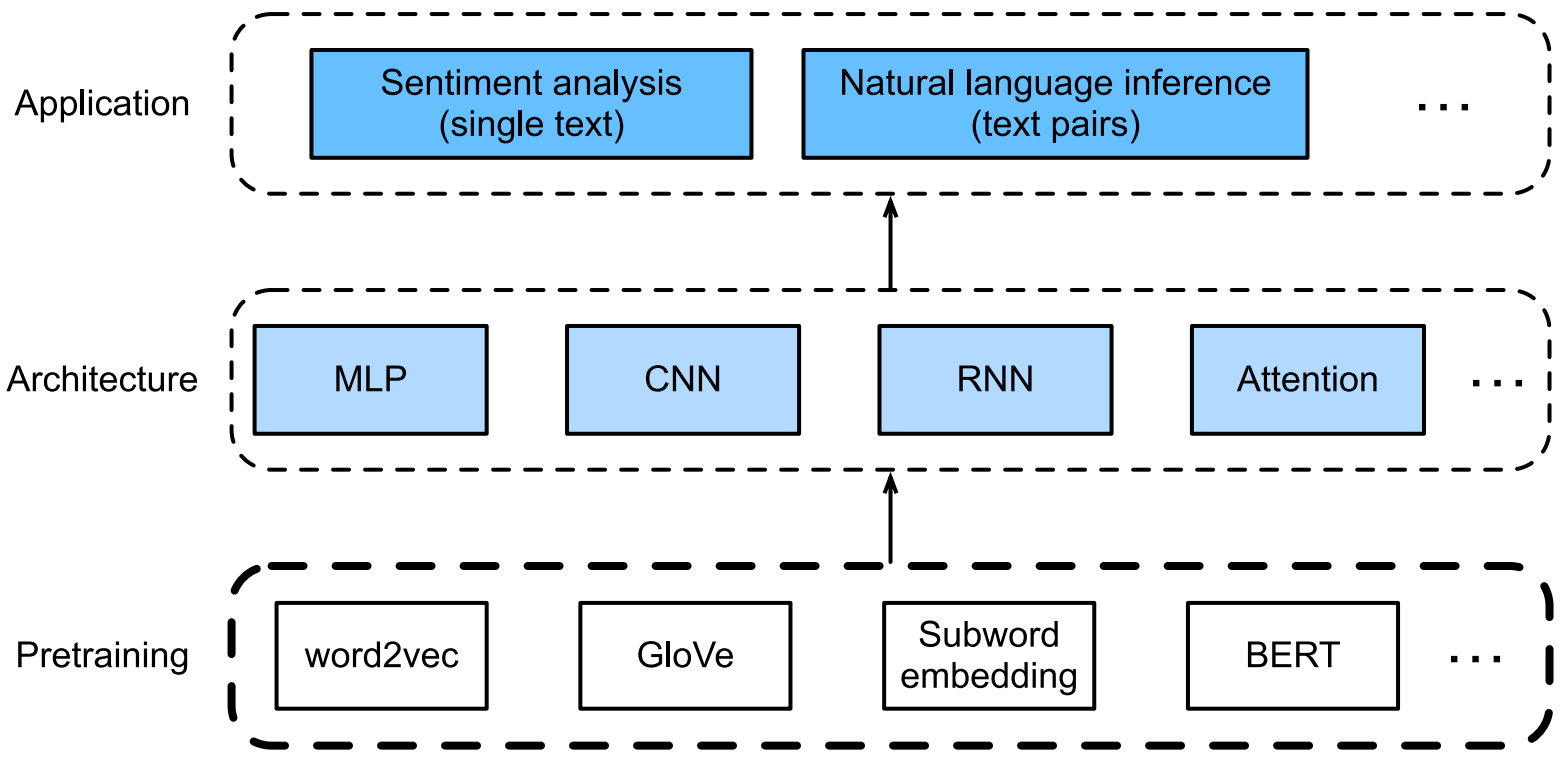

Fig. 15.1 Pretrained text representations can be fed to various deep learning architectures for different downstream natural language processing applications. This chapter focuses on the upstream text representation pretraining.

# 15.1 Word Embedding (word2vec)

Natural language is a complex system used to express meanings. In this system, words are the basic unit of the meaning. As the name implies, *word vectors* are vectors used to represent words, and can also be considered as feature vectors or representations of words. The technique of mapping words to real vectors is called *word embedding*. In recent years, word embedding has gradually become the basic knowledge of natural language processing.

## 15.1.1 One-Hot Vectors Are a Bad Choice

We used one-hot vectors to represent words (characters are words) in Section 9.5. Suppose that the number of different words in the dictionary (the dictionary size) is $N$, and each word corresponds to a different integer (index) from 0 to $N - 1$. To obtain the one-hot vector representation for any word with index $i$, we create a length-$N$ vector with all 0s and set the element at position $i$ to 1. In this way, each word is represented as a vector of length $N$, and it can be used directly by neural networks.

Although one-hot word vectors are easy to construct, they are usually not a good choice. A main reason is that one-hot word vectors cannot accurately express the similarity between different words, such as the *cosine similarity* that we often use. For vectors $\mathbf{x}, \mathbf{y} \in \mathbb{R}^d$, their cosine similarity is the cosine of the angle between them:

$$\frac{\mathbf{x}^\top \mathbf{y}}{\|\mathbf{x}\| \|\mathbf{y}\|} \in [-1, 1]. \tag{15.1.1}$$

Since the cosine similarity between one-hot vectors of any two different words is 0, one-hot vectors cannot encode similarities among words.

## 15.1.2 Self-Supervised word2vec

The word2vec[227] tool was proposed to address the above issue. It maps each word to a fixed-length vector, and these vectors can better express the similarity and analogy relationship among different words. The word2vec tool contains two models, namely *skip-gram* (Mikolov *et al.*, 2013) and *continuous bag of words* (CBOW) (Mikolov *et al.*, 2013). For semantically meaningful representations, their training relies on conditional probabilities that can be viewed as predicting some words using some of their surrounding words in corpora. Since supervision comes from the data without labels, both skip-gram and continuous bag of words are self-supervised models.

In the following, we will introduce these two models and their training methods.

## 15.1.3 The Skip-Gram Model

The *skip-gram* model assumes that a word can be used to generate its surrounding words in a text sequence. Take the text sequence "the", "man", "loves", "his", "son" as an example. Let's choose "loves" as the *center word* and set the context window size to 2. As shown in Fig. 15.1.1, given the center word "loves", the skip-gram model considers the conditional probability for generating the *context words*: "the", "man", "his", and "son", which are no more than 2 words away from the center word:

$$P(\text{"the"}, \text{"man"}, \text{"his"}, \text{"son"} \mid \text{"loves"}). \tag{15.1.2}$$

Assume that the context words are independently generated given the center word (i.e., conditional independence). In this case, the above conditional probability can be rewritten as

$$P(\text{"the"} \mid \text{"loves"}) \cdot P(\text{"man"} \mid \text{"loves"}) \cdot P(\text{"his"} \mid \text{"loves"}) \cdot P(\text{"son"} \mid \text{"loves"}). \tag{15.1.3}$$

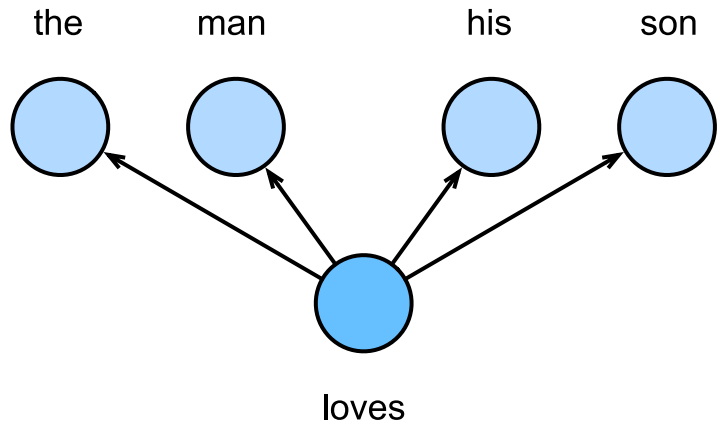

Fig. 15.1.1 The skip-gram model considers the conditional probability of generating the surrounding context words given a center word.

In the skip-gram model, each word has two $d$-dimensional-vector representations for calculating conditional probabilities. More concretely, for any word with index $i$ in the dictionary, denote by $\mathbf{v}_i \in \mathbb{R}^d$ and $\mathbf{u}_i \in \mathbb{R}^d$ its two vectors when used as a *center* word and a *context* word, respectively. The conditional probability of generating any context word $w_o$ (with index $o$ in the dictionary) given the center word $w_c$ (with index $c$ in the dictionary) can be modeled by a softmax operation on vector dot products:

$$P(w_o \mid w_c) = \frac{\exp(\mathbf{u}_o^\top \mathbf{v}_c)}{\sum_{i \in \mathcal{V}} \exp(\mathbf{u}_i^\top \mathbf{v}_c)}, \tag{15.1.4}$$

where the vocabulary index set $\mathcal{V} = \{0, 1, \ldots, |\mathcal{V}| - 1\}$. Given a text sequence of length $T$, where the word at time step $t$ is denoted as $w^{(t)}$. Assume that context words are independently generated given any center word. For context window size $m$, the likelihood function of the skip-gram model is the probability of generating all context words given any center word:

$$\prod_{t=1}^{T} \prod_{-m \leq j \leq m,\ j \neq 0} P(w^{(t+j)} \mid w^{(t)}), \tag{15.1.5}$$

where any time step that is less than 1 or greater than $T$ can be omitted.

## Training

The skip-gram model parameters are the center word vector and context word vector for each word in the vocabulary. In training, we learn the model parameters by maximizing the likelihood function (i.e., maximum likelihood estimation). This is equivalent to minimizing the following loss function:

$$-\sum_{t=1}^{T} \sum_{-m \leq j \leq m,\ j \neq 0} \log P(w^{(t+j)} \mid w^{(t)}). \tag{15.1.6}$$

When using stochastic gradient descent to minimize the loss, in each iteration we can randomly sample a shorter subsequence to calculate the (stochastic) gradient for this subsequence to update the model parameters. To calculate this (stochastic) gradient, we need to obtain the gradients of the log conditional probability with respect to the center word vector and the context word vector. In general, according to (15.1.4) the log conditional probability involving any pair of the center word $w_c$ and the context word $w_o$ is

$$\log P(w_o \mid w_c) = \mathbf{u}_o^\top \mathbf{v}_c - \log\left(\sum_{i \in \mathcal{V}} \exp(\mathbf{u}_i^\top \mathbf{v}_c)\right). \tag{15.1.7}$$

Through differentiation, we can obtain its gradient with respect to the center word vector $\mathbf{v}_c$ as

$$\begin{aligned}
\frac{\partial \log P(w_o \mid w_c)}{\partial \mathbf{v}_c} &= \mathbf{u}_o - \frac{\sum_{j \in \mathcal{V}} \exp(\mathbf{u}_j^\top \mathbf{v}_c)\mathbf{u}_j}{\sum_{i \in \mathcal{V}} \exp(\mathbf{u}_i^\top \mathbf{v}_c)}\\
&= \mathbf{u}_o - \sum_{j \in \mathcal{V}} \left(\frac{\exp(\mathbf{u}_j^\top \mathbf{v}_c)}{\sum_{i \in \mathcal{V}} \exp(\mathbf{u}_i^\top \mathbf{v}_c)}\right) \mathbf{u}_j\\
&= \mathbf{u}_o - \sum_{j \in \mathcal{V}} P(w_j \mid w_c) \mathbf{u}_j.
\end{aligned} \tag{15.1.8}$$

Note that the calculation in (15.1.8) requires the conditional probabilities of all words in the dictionary with $w_c$ as the center word. The gradients for the other word vectors can be obtained in the same way.

After training, for any word with index $i$ in the dictionary, we obtain both word vectors $\mathbf{v}_i$ (as the center word) and $\mathbf{u}_i$ (as the context word). In natural language processing applications, the center word vectors of the skip-gram model are typically used as the word representations.

### 15.1.4 The Continuous Bag of Words (CBOW) Model

The *continuous bag of words* (CBOW) model is similar to the skip-gram model. The major difference from the skip-gram model is that the continuous bag of words model assumes that a center word is generated based on its surrounding context words in the text sequence. For example, in the same text sequence "the", "man", "loves", "his", and "son", with "loves" as the center word and the context window size being 2, the continuous bag of words model considers the conditional probability of generating the center word "loves" based on the context words "the", "man", "his" and "son" (as shown in Fig. 15.1.2), which is

$$P(\text{"loves"} \mid \text{"the"}, \text{"man"}, \text{"his"}, \text{"son"}). \tag{15.1.9}$$

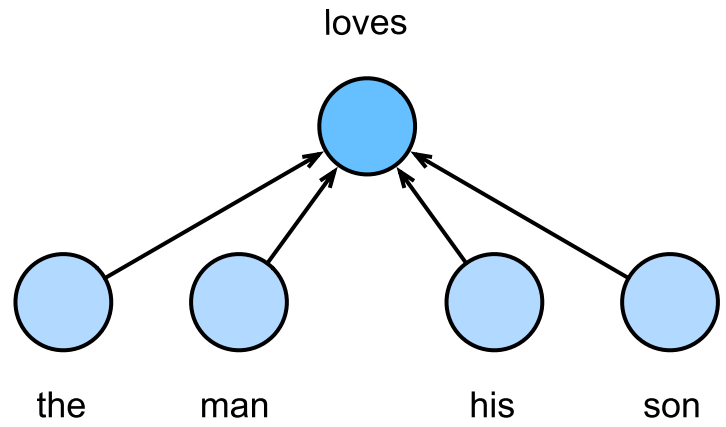

Fig. 15.1.2 The continuous bag of words model considers the conditional probability of generating the center word given its surrounding context words.

Since there are multiple context words in the continuous bag of words model, these context word vectors are averaged in the calculation of the conditional probability. Specifically, for any word with index $i$ in the dictionary, denote by $\mathbf{v}_i \in \mathbb{R}^d$ and $\mathbf{u}_i \in \mathbb{R}^d$ its two vectors when used as a *context* word and a *center* word (meanings are switched in the skip-gram model), respectively. The conditional probability of generating any center word $w_c$ (with index $c$ in the dictionary) given its surrounding context words $w_{o_1}, \ldots, w_{o_{2m}}$ (with index $o_1, \ldots, o_{2m}$ in the dictionary) can be modeled by

$$P(w_c \mid w_{o_1}, \ldots, w_{o_{2m}}) = \frac{\exp\left(\frac{1}{2m}\mathbf{u}_c^\top (\mathbf{v}_{o_1} + \ldots + \mathbf{v}_{o_{2m}})\right)}{\sum_{i \in \mathcal{V}} \exp\left(\frac{1}{2m}\mathbf{u}_i^\top (\mathbf{v}_{o_1} + \ldots + \mathbf{v}_{o_{2m}})\right)}. \tag{15.1.10}$$

For brevity, let $\mathcal{W}_o = \{w_{o_1}, \ldots, w_{o_{2m}}\}$ and $\bar{\mathbf{v}}_o = \left(\mathbf{v}_{o_1} + \ldots + \mathbf{v}_{o_{2m}}\right)/(2m)$. Then (15.1.10)

can be simplified as

$$P(w_c \mid \mathcal{W}_o) = \frac{\exp\left(\mathbf{u}_c^\top \bar{\mathbf{v}}_o\right)}{\sum_{i \in \mathcal{V}} \exp\left(\mathbf{u}_i^\top \bar{\mathbf{v}}_o\right)}. \tag{15.1.11}$$

Given a text sequence of length $T$, where the word at time step $t$ is denoted as $w^{(t)}$. For context window size $m$, the likelihood function of the continuous bag of words model is the probability of generating all center words given their context words:

$$\prod_{t=1}^{T} P(w^{(t)} \mid w^{(t-m)}, \ldots, w^{(t-1)}, w^{(t+1)}, \ldots, w^{(t+m)}). \tag{15.1.12}$$

### Training

Training continuous bag of words models is almost the same as training skip-gram models. The maximum likelihood estimation of the continuous bag of words model is equivalent to minimizing the following loss function:

$$-\sum_{t=1}^{T} \log P(w^{(t)} \mid w^{(t-m)}, \ldots, w^{(t-1)}, w^{(t+1)}, \ldots, w^{(t+m)}). \tag{15.1.13}$$

Notice that

$$\log\, P(w_c \mid \mathcal{W}_o) = \mathbf{u}_c^\top \bar{\mathbf{v}}_o - \log\left(\sum_{i \in \mathcal{V}} \exp\left(\mathbf{u}_i^\top \bar{\mathbf{v}}_o\right)\right). \tag{15.1.14}$$

Through differentiation, we can obtain its gradient with respect to any context word vector $\mathbf{v}_{o_i} (i = 1, \ldots, 2m)$ as

$$\frac{\partial \log\, P(w_c \mid \mathcal{W}_o)}{\partial \mathbf{v}_{o_i}} = \frac{1}{2m}\left(\mathbf{u}_c - \sum_{j \in \mathcal{V}} \frac{\exp(\mathbf{u}_j^\top \bar{\mathbf{v}}_o)\mathbf{u}_j}{\sum_{i \in \mathcal{V}} \exp(\mathbf{u}_i^\top \bar{\mathbf{v}}_o)}\right) = \frac{1}{2m}\left(\mathbf{u}_c - \sum_{j \in \mathcal{V}} P(w_j \mid \mathcal{W}_o)\mathbf{u}_j\right). \tag{15.1.15}$$

The gradients for the other word vectors can be obtained in the same way. Unlike the skip-gram model, the continuous bag of words model typically uses context word vectors as the word representations.

## 15.1.5 Summary

- Word vectors are vectors used to represent words, and can also be considered as feature vectors or representations of words. The technique of mapping words to real vectors is called word embedding.
- The word2vec tool contains both the skip-gram and continuous bag of words models.
- The skip-gram model assumes that a word can be used to generate its surrounding words in a text sequence; while the continuous bag of words model assumes that a center word is generated based on its surrounding context words.

### 15.1.6 Exercises

1. What is the computational complexity for calculating each gradient? What could be the issue if the dictionary size is huge?
2. Some fixed phrases in English consist of multiple words, such as “new york”. How to train their word vectors? Hint: see Section 4 in the word2vec paper (Mikolov *et al.*, 2013).
3. Let’s reflect on the word2vec design by taking the skip-gram model as an example. What is the relationship between the dot product of two word vectors in the skip-gram model and the cosine similarity? For a pair of words with similar semantics, why may the cosine similarity of their word vectors (trained by the skip-gram model) be high?

Discussions[228].

## 15.2 Approximate Training

Recall our discussions in Section 15.1. The main idea of the skip-gram model is using softmax operations to calculate the conditional probability of generating a context word $w_o$ based on the given center word $w_c$ in (15.1.4), whose corresponding logarithmic loss is given by the opposite of (15.1.7).

Due to the nature of the softmax operation, since a context word may be anyone in the dictionary $\mathcal{V}$, the opposite of (15.1.7) contains the summation of items as many as the entire size of the vocabulary. Consequently, the gradient calculation for the skip-gram model in (15.1.8) and that for the continuous bag-of-words model in (15.1.15) both contain the summation. Unfortunately, the computational cost for such gradients that sum over a large dictionary (often with hundreds of thousands or millions of words) is huge!

In order to reduce the aforementioned computational complexity, this section will introduce two approximate training methods: *negative sampling* and *hierarchical softmax*. Due to the similarity between the skip-gram model and the continuous bag of words model, we will just take the skip-gram model as an example to describe these two approximate training methods.

### 15.2.1 Negative Sampling

Negative sampling modifies the original objective function. Given the context window of a center word $w_c$, the fact that any (context) word $w_o$ comes from this context window is considered as an event with the probability modeled by

$$P(D = 1 \mid w_c, w_o) = \sigma(\mathbf{u}_o^\top \mathbf{v}_c), \tag{15.2.1}$$

where $\sigma$ uses the definition of the sigmoid activation function:

$$\sigma(x) = \frac{1}{1 + \exp(-x)}. \tag{15.2.2}$$

Let's begin by maximizing the joint probability of all such events in text sequences to train word embeddings. Specifically, given a text sequence of length $T$, denote by $w^{(t)}$ the word at time step $t$ and let the context window size be $m$, consider maximizing the joint probability

$$\prod_{t=1}^{T} \prod_{-m \leq j \leq m,\ j \neq 0} P(D = 1 \mid w^{(t)}, w^{(t+j)}). \tag{15.2.3}$$

However, (15.2.3) only considers those events that involve positive examples. As a result, the joint probability in (15.2.3) is maximized to 1 only if all the word vectors are equal to infinity. Of course, such results are meaningless. To make the objective function more meaningful, *negative sampling* adds negative examples sampled from a predefined distribution.

Denote by $S$ the event that a context word $w_o$ comes from the context window of a center word $w_c$. For this event involving $w_o$, from a predefined distribution $P(w)$ sample $K$ *noise words* that are not from this context window. Denote by $N_k$ the event that a noise word $w_k$ ($k = 1, \ldots, K$) does not come from the context window of $w_c$. Assume that these events involving both the positive example and negative examples $S, N_1, \ldots, N_K$ are mutually independent. Negative sampling rewrites the joint probability (involving only positive examples) in (15.2.3) as

$$\prod_{t=1}^{T} \prod_{-m \leq j \leq m,\ j \neq 0} P(w^{(t+j)} \mid w^{(t)}), \tag{15.2.4}$$

where the conditional probability is approximated through events $S, N_1, \ldots, N_K$:

$$P(w^{(t+j)} \mid w^{(t)}) = P(D = 1 \mid w^{(t)}, w^{(t+j)}) \prod_{k=1,\ w_k \sim P(w)}^{K} P(D = 0 \mid w^{(t)}, w_k). \tag{15.2.5}$$

Denote by $i_t$ and $h_k$ the indices of a word $w^{(t)}$ at time step $t$ of a text sequence and a noise word $w_k$, respectively. The logarithmic loss with respect to the conditional probabilities in (15.2.5) is

$$\begin{aligned}
-\log P(w^{(t+j)} \mid w^{(t)}) &= -\log P(D = 1 \mid w^{(t)}, w^{(t+j)}) - \sum_{k=1,\ w_k \sim P(w)}^{K} \log P(D = 0 \mid w^{(t)}, w_k) \\
&= -\log \sigma\left(\mathbf{u}_{i_{t+j}}^\top \mathbf{v}_{i_t}\right) - \sum_{k=1,\ w_k \sim P(w)}^{K} \log\left(1 - \sigma\left(\mathbf{u}_{h_k}^\top \mathbf{v}_{i_t}\right)\right) \\
&= -\log \sigma\left(\mathbf{u}_{i_{t+j}}^\top \mathbf{v}_{i_t}\right) - \sum_{k=1,\ w_k \sim P(w)}^{K} \log \sigma\left(-\mathbf{u}_{h_k}^\top \mathbf{v}_{i_t}\right).
\end{aligned} \tag{15.2.6}$$

We can see that now the computational cost for gradients at each training step has nothing to do with the dictionary size, but linearly depends on $K$. When setting the hyperparameter $K$ to a smaller value, the computational cost for gradients at each training step with negative sampling is smaller.

### 15.2.2 Hierarchical Softmax

As an alternative approximate training method, *hierarchical softmax* uses the binary tree, a data structure illustrated in Fig. 15.2.1, where each leaf node of the tree represents a word in dictionary $\mathcal{V}$.

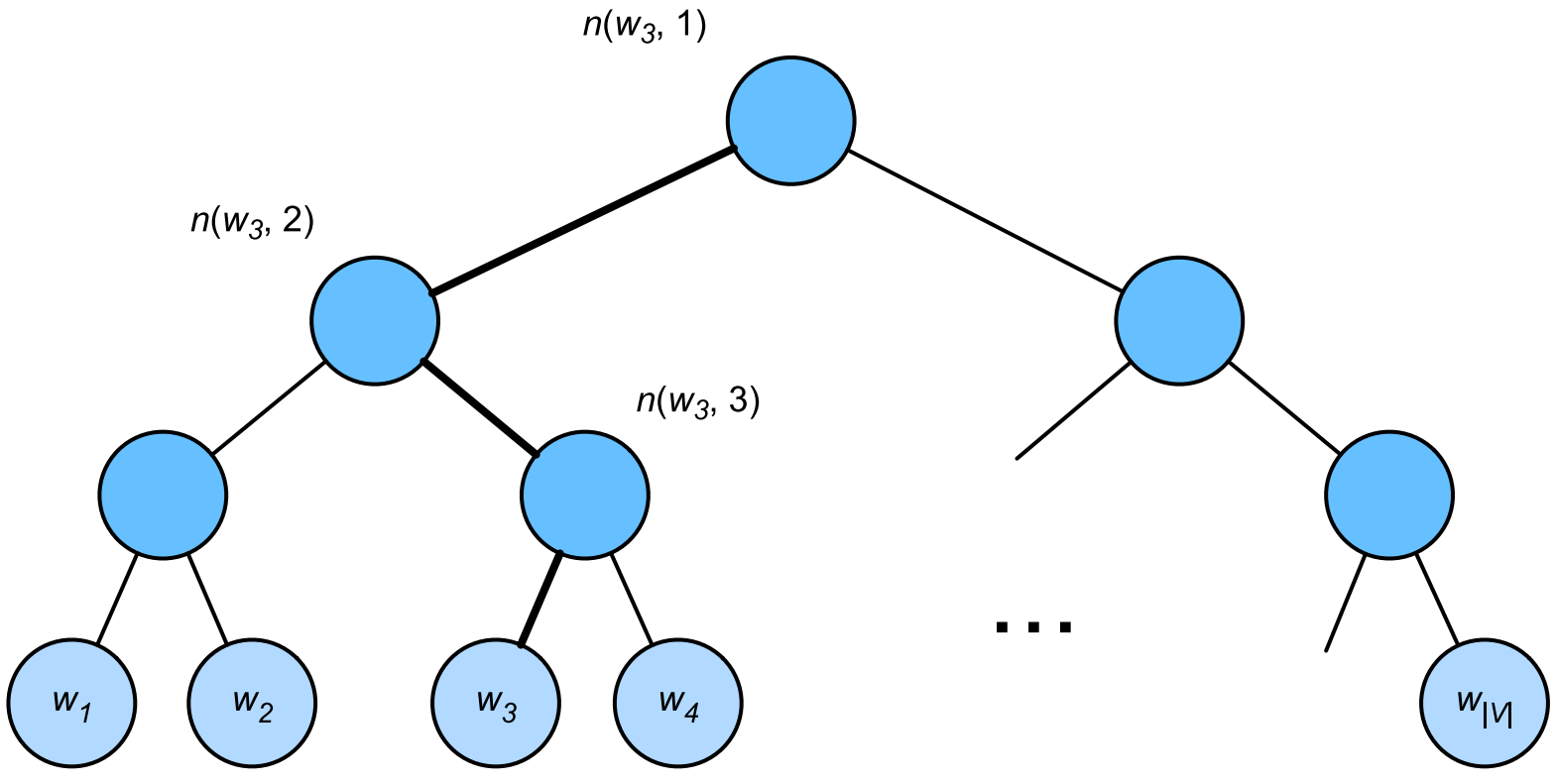

Fig. 15.2.1 Hierarchical softmax for approximate training, where each leaf node of the tree represents a word in the dictionary.

Denote by $L(w)$ the number of nodes (including both ends) on the path from the root node to the leaf node representing word $w$ in the binary tree. Let $n(w, j)$ be the $j^{\textrm{th}}$ node on this path, with its context word vector being $\mathbf{u}_{n(w,j)}$. For example, $L(w_3) = 4$ in Fig. 15.2.1. Hierarchical softmax approximates the conditional probability in (15.1.4) as

$$P(w_o \mid w_c) = \prod_{j=1}^{L(w_o)-1} \sigma\left( [\![ n(w_o, j+1) = \textrm{leftChild}(n(w_o, j)) ]\!] \cdot \mathbf{u}_{n(w_o, j)}^\top \mathbf{v}_c\right), \quad (15.2.7)$$

where function $\sigma$ is defined in (15.2.2), and $\textrm{leftChild}(n)$ is the left child node of node $n$: if $x$ is true, $[\![x]\!] = 1$; otherwise $[\![x]\!] = -1$.

To illustrate, let's calculate the conditional probability of generating word $w_3$ given word $w_c$ in Fig. 15.2.1. This requires dot products between the word vector $\mathbf{v}_c$ of $w_c$ and non-leaf node vectors on the path (the path in bold in Fig. 15.2.1) from the root to $w_3$, which is traversed left, right, then left:

$$P(w_3 \mid w_c) = \sigma(\mathbf{u}_{n(w_3, 1)}^\top \mathbf{v}_c) \cdot \sigma(-\mathbf{u}_{n(w_3, 2)}^\top \mathbf{v}_c) \cdot \sigma(\mathbf{u}_{n(w_3, 3)}^\top \mathbf{v}_c). \quad (15.2.8)$$

Since $\sigma(x)+ \sigma(-x) = 1$, it holds that the conditional probabilities of generating all the words in dictionary $\mathcal{V}$ based on any word $w_c$ sum up to one:

$$\sum_{w \in \mathcal{V}} P(w \mid w_c) = 1. \quad (15.2.9)$$

Fortunately, since $L(w_o) - 1$ is on the order of $\mathcal{O}(\log_2|\mathcal{V}|)$ due to the binary tree structure, when the dictionary size $\mathcal{V}$ is huge, the computational cost for each training step using hierarchical softmax is significantly reduced compared with that without approximate training.

### 15.2.3 Summary

- Negative sampling constructs the loss function by considering mutually independent events that involve both positive and negative examples. The computational cost for training is linearly dependent on the number of noise words at each step.
- Hierarchical softmax constructs the loss function using the path from the root node to the leaf node in the binary tree. The computational cost for training is dependent on the logarithm of the dictionary size at each step.

### 15.2.4 Exercises

1. How can we sample noise words in negative sampling?
2. Verify that (15.2.9) holds.
3. How to train the continuous bag of words model using negative sampling and hierarchical softmax, respectively?

Discussions[229].

## 15.3 The Dataset for Pretraining Word Embeddings

Now that we know the technical details of the word2vec models and approximate training methods, let's walk through their implementations. Specifically, we will take the skip-gram model in Section 15.1 and negative sampling in Section 15.2 as an example. In this section, we begin with the dataset for pretraining the word embedding model: the original format of the data will be transformed into minibatches that can be iterated over during training.

```
import collections
import math
import os
import random
import torch
from d2l import torch as d2l
```

## 15.3.1 Reading the Dataset

The dataset that we use here is Penn Tree Bank (PTB)[230]. This corpus is sampled from Wall Street Journal articles, split into training, validation, and test sets. In the original format, each line of the text file represents a sentence of words that are separated by spaces. Here we treat each word as a token.

```
#@save
d2l.DATA_HUB['ptb'] = (d2l.DATA_URL + 'ptb.zip',
                       '319d85e578af0cdc590547f26231e4e31cdf1e42')

#@save
def read_ptb():
    """Load the PTB dataset into a list of text lines."""
    data_dir = d2l.download_extract('ptb')
    # Read the training set
    with open(os.path.join(data_dir, 'ptb.train.txt')) as f:
        raw_text = f.read()
    return [line.split() for line in raw_text.split('\n')]

sentences = read_ptb()
f'# sentences: {len(sentences)}'
```

```
'# sentences: 42069'
```

After reading the training set, we build a vocabulary for the corpus, where any word that appears less than 10 times is replaced by the "<unk>" token. Note that the original dataset also contains "<unk>" tokens that represent rare (unknown) words.

```
vocab = d2l.Vocab(sentences, min_freq=10)
f'vocab size: {len(vocab)}'
```

```
'vocab size: 6719'
```

## 15.3.2 Subsampling

Text data typically have high-frequency words such as "the", "a", and "in": they may even occur billions of times in very large corpora. However, these words often co-occur with many different words in context windows, providing little useful signals. For instance, consider the word "chip" in a context window: intuitively its co-occurrence with a low-frequency word "intel" is more useful in training than the co-occurrence with a high-frequency word "a". Moreover, training with vast amounts of (high-frequency) words is slow. Thus, when training word embedding models, high-frequency words can be *subsampled* (Mikolov *et al.*, 2013). Specifically, each indexed word $w_i$ in the dataset will be discarded with probability

$$P(w_i) = \max\left(1 - \sqrt{\frac{t}{f(w_i)}}, 0\right), \tag{15.3.1}$$

where $f(w_i)$ is the ratio of the number of words $w_i$ to the total number of words in the dataset, and the constant $t$ is a hyperparameter ($10^{-4}$ in the experiment). We can see that only when the relative frequency $f(w_i) > t$ can the (high-frequency) word $w_i$ be discarded, and the higher the relative frequency of the word, the greater the probability of being discarded.

```
#@save
def subsample(sentences, vocab):
    """Subsample high-frequency words."""
    # Exclude unknown tokens ('<unk>')
    sentences = [[token for token in line if vocab[token] != vocab.unk]
                 for line in sentences]
    counter = collections.Counter([
        token for line in sentences for token in line])
    num_tokens = sum(counter.values())

    # Return True if `token` is kept during subsampling
    def keep(token):
        return(random.uniform(0, 1) <
               math.sqrt(1e-4 / counter[token] * num_tokens))

    return ([[token for token in line if keep(token)] for line in sentences],
            counter)

subsampled, counter = subsample(sentences, vocab)
```

The following code snippet plots the histogram of the number of tokens per sentence before and after subsampling. As expected, subsampling significantly shortens sentences by dropping high-frequency words, which will lead to training speedup.

```
d2l.show_list_len_pair_hist(['origin', 'subsampled'], '# tokens per sentence',
                            'count', sentences, subsampled);
```

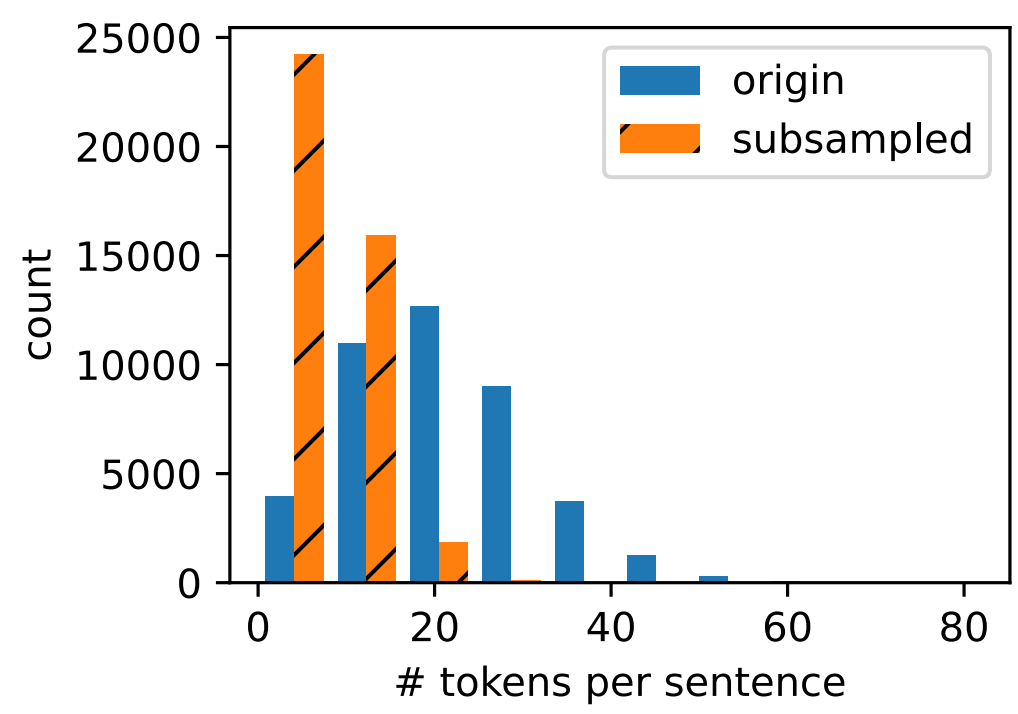

For individual tokens, the sampling rate of the high-frequency word “the” is less than 1/20.

```
def compare_counts(token):
    return (f'# of "{token}": '
```

```
            f'before={sum([l.count(token) for l in sentences])}, '
            f'after={sum([l.count(token) for l in subsampled])}')

compare_counts('the')
```

```
'# of "the": before=50770, after=1992'
```

In contrast, low-frequency words "join" are completely kept.

```
compare_counts('join')
```

```
'# of "join": before=45, after=45'
```

After subsampling, we map tokens to their indices for the corpus.

```
corpus = [vocab[line] for line in subsampled]
corpus[:3]
```

```
[[], [4127, 3228, 710, 1773, 4060], [3895, 3922, 1922, 4743]]
```

### 15.3.3 Extracting Center Words and Context Words

The following `get_centers_and_contexts` function extracts all the center words and their context words from `corpus`. It uniformly samples an integer between 1 and `max_window_size` at random as the context window size. For any center word, those words whose distance from it does not exceed the sampled context window size are its context words.

```
#@save
def get_centers_and_contexts(corpus, max_window_size):
    """Return center words and context words in skip-gram."""
    centers, contexts = [], []
    for line in corpus:
        # To form a "center word--context word" pair, each sentence needs to
        # have at least 2 words
        if len(line) < 2:
            continue
        centers += line
        for i in range(len(line)):  # Context window centered at `i`
            window_size = random.randint(1, max_window_size)
            indices = list(range(max(0, i - window_size),
                                 min(len(line), i + 1 + window_size)))
            # Exclude the center word from the context words
            indices.remove(i)
            contexts.append([line[idx] for idx in indices])
    return centers, contexts
```

Next, we create an artificial dataset containing two sentences of 7 and 3 words, respectively. Let the maximum context window size be 2 and print all the center words and their context words.

```
tiny_dataset = [list(range(7)), list(range(7, 10))]
print('dataset', tiny_dataset)
for center, context in zip(*get_centers_and_contexts(tiny_dataset, 2)):
    print('center', center, 'has contexts', context)
```

```
dataset [[0, 1, 2, 3, 4, 5, 6], [7, 8, 9]]
center 0 has contexts [1]
center 1 has contexts [0, 2]
center 2 has contexts [0, 1, 3, 4]
center 3 has contexts [1, 2, 4, 5]
center 4 has contexts [3, 5]
center 5 has contexts [3, 4, 6]
center 6 has contexts [5]
center 7 has contexts [8, 9]
center 8 has contexts [7, 9]
center 9 has contexts [7, 8]
```

When training on the PTB dataset, we set the maximum context window size to 5. The following extracts all the center words and their context words in the dataset.

```
all_centers, all_contexts = get_centers_and_contexts(corpus, 5)
f'# center-context pairs: {sum([len(contexts) for contexts in all_contexts])}'
```

```
'# center-context pairs: 1503184'
```

### 15.3.4 Negative Sampling

We use negative sampling for approximate training. To sample noise words according to a predefined distribution, we define the following `RandomGenerator` class, where the (possibly unnormalized) sampling distribution is passed via the argument `sampling_weights`.

```
#@save
class RandomGenerator:
    """Randomly draw among {1, ..., n} according to n sampling weights."""
    def __init__(self, sampling_weights):
        # Exclude
        self.population = list(range(1, len(sampling_weights) + 1))
        self.sampling_weights = sampling_weights
        self.candidates = []
        self.i = 0

    def draw(self):
        if self.i == len(self.candidates):
            # Cache `k` random sampling results
```

```
            self.candidates = random.choices(
                self.population, self.sampling_weights, k=10000)
            self.i = 0
        self.i += 1
        return self.candidates[self.i - 1]
```

For example, we can draw 10 random variables $X$ among indices 1, 2, and 3 with sampling probabilities $P(X = 1) = 2/9, P(X = 2) = 3/9$, and $P(X = 3) = 4/9$ as follows.

For a pair of center word and context word, we randomly sample `K` (5 in the experiment) noise words. According to the suggestions in the word2vec paper, the sampling probability $P(w)$ of a noise word $w$ is set to its relative frequency in the dictionary raised to the power of 0.75 (Mikolov *et al.*, 2013).

```
#@save
def get_negatives(all_contexts, vocab, counter, K):
    """Return noise words in negative sampling."""
    # Sampling weights for words with indices 1, 2, ... (index 0 is the
    # excluded unknown token) in the vocabulary
    sampling_weights = [counter[vocab.to_tokens(i)]**0.75
                        for i in range(1, len(vocab))]
    all_negatives, generator = [], RandomGenerator(sampling_weights)
    for contexts in all_contexts:
        negatives = []
        while len(negatives) < len(contexts) * K:
            neg = generator.draw()
            # Noise words cannot be context words
            if neg not in contexts:
                negatives.append(neg)
        all_negatives.append(negatives)
    return all_negatives

all_negatives = get_negatives(all_contexts, vocab, counter, 5)
```

### 15.3.5 Loading Training Examples in Minibatches

After all the center words together with their context words and sampled noise words are extracted, they will be transformed into minibatches of examples that can be iteratively loaded during training.

In a minibatch, the $i^{\text{th}}$ example includes a center word and its $n_i$ context words and $m_i$ noise words. Due to varying context window sizes, $n_i + m_i$ varies for different $i$. Thus, for each example we concatenate its context words and noise words in the `contexts_negatives` variable, and pad zeros until the concatenation length reaches $\max_i n_i + m_i$ (`max_len`). To exclude paddings in the calculation of the loss, we define a mask variable `masks`. There is a one-to-one correspondence between elements in `masks` and elements in `contexts_negatives`, where zeros (otherwise ones) in `masks` correspond to paddings in `contexts_negatives`.

To distinguish between positive and negative examples, we separate context words from noise words in `contexts_negatives` via a `labels` variable. Similar to `masks`, there is also a one-to-one correspondence between elements in `labels` and elements in `contexts_negatives`, where ones (otherwise zeros) in `labels` correspond to context words (positive examples) in `contexts_negatives`.

The above idea is implemented in the following `batchify` function. Its input `data` is a list with length equal to the batch size, where each element is an example consisting of the center word `center`, its context words `context`, and its noise words `negative`. This function returns a minibatch that can be loaded for calculations during training, such as including the mask variable.

```
#@save
def batchify(data):
    """Return a minibatch of examples for skip-gram with negative sampling."""
    max_len = max(len(c) + len(n) for _, c, n in data)
    centers, contexts_negatives, masks, labels = [], [], [], []
    for center, context, negative in data:
        cur_len = len(context) + len(negative)
        centers += [center]
        contexts_negatives += [context + negative + [0] * (max_len - cur_len)]
        masks += [[1] * cur_len + [0] * (max_len - cur_len)]
        labels += [[1] * len(context) + [0] * (max_len - len(context))]
    return (torch.tensor(centers).reshape((-1, 1)), torch.tensor(
        contexts_negatives), torch.tensor(masks), torch.tensor(labels))
```

Let's test this function using a minibatch of two examples.

```
x_1 = (1, [2, 2], [3, 3, 3, 3])
x_2 = (1, [2, 2, 2], [3, 3])
batch = batchify((x_1, x_2))

names = ['centers', 'contexts_negatives', 'masks', 'labels']
for name, data in zip(names, batch):
    print(name, '=', data)
```

```
centers = tensor([[1],
        [1]])
contexts_negatives = tensor([[2, 2, 3, 3, 3, 3],
        [2, 2, 2, 3, 3, 0]])
masks = tensor([[1, 1, 1, 1, 1, 1],
        [1, 1, 1, 1, 1, 0]])
labels = tensor([[1, 1, 0, 0, 0, 0],
        [1, 1, 1, 0, 0, 0]])
```

## 15.3.6 Putting It All Together

Last, we define the `load_data_ptb` function that reads the PTB dataset and returns the data iterator and the vocabulary.

```
#@save
def load_data_ptb(batch_size, max_window_size, num_noise_words):
    """Download the PTB dataset and then load it into memory."""
    num_workers = d2l.get_dataloader_workers()
    sentences = read_ptb()
    vocab = d2l.Vocab(sentences, min_freq=10)
    subsampled, counter = subsample(sentences, vocab)
    corpus = [vocab[line] for line in subsampled]
    all_centers, all_contexts = get_centers_and_contexts(
        corpus, max_window_size)
    all_negatives = get_negatives(
        all_contexts, vocab, counter, num_noise_words)

    class PTBDataset(torch.utils.data.Dataset):
        def __init__(self, centers, contexts, negatives):
            assert len(centers) == len(contexts) == len(negatives)
            self.centers = centers
            self.contexts = contexts
            self.negatives = negatives

        def __getitem__(self, index):
            return (self.centers[index], self.contexts[index],
                    self.negatives[index])

        def __len__(self):
            return len(self.centers)

    dataset = PTBDataset(all_centers, all_contexts, all_negatives)

    data_iter = torch.utils.data.DataLoader(dataset, batch_size, shuffle=True,
                                      collate_fn=batchify,
                                      num_workers=num_workers)
    return data_iter, vocab
```

Let's print the first minibatch of the data iterator.

```
data_iter, vocab = load_data_ptb(512, 5, 5)
for batch in data_iter:
    for name, data in zip(names, batch):
        print(name, 'shape:', data.shape)
    break
```

```
centers shape: torch.Size([512, 1])
contexts_negatives shape: torch.Size([512, 60])
masks shape: torch.Size([512, 60])
labels shape: torch.Size([512, 60])
```

### 15.3.7 Summary

- High-frequency words may not be so useful in training. We can subsample them for speedup in training.

- For computational efficiency, we load examples in minibatches. We can define other variables to distinguish paddings from non-paddings, and positive examples from negative ones.

### 15.3.8 Exercises

1. How does the running time of code in this section changes if not using subsampling?
2. The `RandomGenerator` class caches `k` random sampling results. Set `k` to other values and see how it affects the data loading speed.
3. What other hyperparameters in the code of this section may affect the data loading speed?

Discussions[231].

## 15.4 Pretraining word2vec

We go on to implement the skip-gram model defined in Section 15.1. Then we will pretrain word2vec using negative sampling on the PTB dataset. First of all, let's obtain the data iterator and the vocabulary for this dataset by calling the `d2l.load_data_ptb` function, which was described in Section 15.3

```
import math
import torch
from torch import nn
from d2l import torch as d2l

batch_size, max_window_size, num_noise_words = 512, 5, 5
data_iter, vocab = d2l.load_data_ptb(batch_size, max_window_size,
                                     num_noise_words)
```

### 15.4.1 The Skip-Gram Model

We implement the skip-gram model by using embedding layers and batch matrix multiplications. First, let's review how embedding layers work.

#### Embedding Layer

As described in Section 10.7, an embedding layer maps a token's index to its feature vector. The weight of this layer is a matrix whose number of rows equals to the dictionary size (`input_dim`) and number of columns equals to the vector dimension for each token (`output_dim`). After a word embedding model is trained, this weight is what we need.

```
embed = nn.Embedding(num_embeddings=20, embedding_dim=4)
print(f'Parameter embedding_weight ({embed.weight.shape}, '
      f'dtype={embed.weight.dtype})')
```

```
Parameter embedding_weight (torch.Size([20, 4]), dtype=torch.float32)
```

The input of an embedding layer is the index of a token (word). For any token index $i$, its vector representation can be obtained from the $i^{\text{th}}$ row of the weight matrix in the embedding layer. Since the vector dimension (`output_dim`) was set to 4, the embedding layer returns vectors with shape (2, 3, 4) for a minibatch of token indices with shape (2, 3).

```
x = torch.tensor([[1, 2, 3], [4, 5, 6]])
embed(x)
```

```
tensor([[[ 1.4289,  0.6862,  1.3745, -0.0402],
         [ 0.0612, -0.4935,  0.7515,  1.2910],
         [ 1.2522,  0.1610, -1.4299, -0.7257]],

        [[ 0.1162,  0.2630, -2.0102, -0.9138],
         [-0.9011,  0.4052,  0.4618,  0.5271],
         [-1.8482,  0.5460,  0.0158, -0.2720]]], grad_fn=<EmbeddingBackward0>)
```

### Defining the Forward Propagation

In the forward propagation, the input of the skip-gram model includes the center word indices `center` of shape (batch size, 1) and the concatenated context and noise word indices `contexts_and_negatives` of shape (batch size, `max_len`), where `max_len` is defined in Section 15.3.5. These two variables are first transformed from the token indices into vectors via the embedding layer, then their batch matrix multiplication (described in Section 11.3.2) returns an output of shape (batch size, 1, `max_len`). Each element in the output is the dot product of a center word vector and a context or noise word vector.

```
def skip_gram(center, contexts_and_negatives, embed_v, embed_u):
    v = embed_v(center)
    u = embed_u(contexts_and_negatives)
    pred = torch.bmm(v, u.permute(0, 2, 1))
    return pred
```

Let's print the output shape of this `skip_gram` function for some example inputs.

```
skip_gram(torch.ones((2, 1), dtype=torch.long),
          torch.ones((2, 4), dtype=torch.long), embed, embed).shape
```

```
torch.Size([2, 1, 4])
```

## 15.4.2 Training

Before training the skip-gram model with negative sampling, let's first define its loss function.

### Binary Cross-Entropy Loss

According to the definition of the loss function for negative sampling in Section 15.2.1, we will use the binary cross-entropy loss.

```
class SigmoidBCELoss(nn.Module):
    # Binary cross-entropy loss with masking
    def __init__(self):
        super().__init__()

    def forward(self, inputs, target, mask=None):
        out = nn.functional.binary_cross_entropy_with_logits(
            inputs, target, weight=mask, reduction="none")
        return out.mean(dim=1)

loss = SigmoidBCELoss()
```

Recall our descriptions of the mask variable and the label variable in Section 15.3.5. The following calculates the binary cross-entropy loss for the given variables.

```
pred = torch.tensor([[1.1, -2.2, 3.3, -4.4]] * 2)
label = torch.tensor([[1.0, 0.0, 0.0, 0.0], [0.0, 1.0, 0.0, 0.0]])
mask = torch.tensor([[1, 1, 1, 1], [1, 1, 0, 0]])
loss(pred, label, mask) * mask.shape[1] / mask.sum(axis=1)
```

```
tensor([0.9352, 1.8462])
```

Below shows how the above results are calculated (in a less efficient way) using the sigmoid activation function in the binary cross-entropy loss. We can consider the two outputs as two normalized losses that are averaged over non-masked predictions.

```
def sigmd(x):
    return -math.log(1 / (1 + math.exp(-x)))

print(f'{(sigmd(1.1) + sigmd(2.2) + sigmd(-3.3) + sigmd(4.4)) / 4:.4f}')
print(f'{(sigmd(-1.1) + sigmd(-2.2)) / 2:.4f}')
```

```
0.9352
1.8462
```

### Initializing Model Parameters

We define two embedding layers for all the words in the vocabulary when they are used as center words and context words, respectively. The word vector dimension `embed_size` is set to 100.

```
embed_size = 100
net = nn.Sequential(nn.Embedding(num_embeddings=len(vocab),
                                 embedding_dim=embed_size),
                    nn.Embedding(num_embeddings=len(vocab),
                                 embedding_dim=embed_size))
```

### Defining the Training Loop

The training loop is defined below. Because of the existence of padding, the calculation of the loss function is slightly different compared to the previous training functions.

```
def train(net, data_iter, lr, num_epochs, device=d2l.try_gpu()):
    def init_weights(module):
        if type(module) == nn.Embedding:
            nn.init.xavier_uniform_(module.weight)
    net.apply(init_weights)
    net = net.to(device)
    optimizer = torch.optim.Adam(net.parameters(), lr=lr)
    animator = d2l.Animator(xlabel='epoch', ylabel='loss',
                            xlim=[1, num_epochs])
    # Sum of normalized losses, no. of normalized losses
    metric = d2l.Accumulator(2)
    for epoch in range(num_epochs):
        timer, num_batches = d2l.Timer(), len(data_iter)
        for i, batch in enumerate(data_iter):
            optimizer.zero_grad()
            center, context_negative, mask, label = [
                data.to(device) for data in batch]

            pred = skip_gram(center, context_negative, net[0], net[1])
            l = (loss(pred.reshape(label.shape).float(), label.float(), mask)
                     / mask.sum(axis=1) * mask.shape[1])
            l.sum().backward()
            optimizer.step()
            metric.add(l.sum(), l.numel())
            if (i + 1) % (num_batches // 5) == 0 or i == num_batches - 1:
                animator.add(epoch + (i + 1) / num_batches,
                             (metric[0] / metric[1],))
```

```
    print(f'loss {metric[0] / metric[1]:.3f}, '
          f'{metric[1] / timer.stop():.1f} tokens/sec on {str(device)}')
```

Now we can train a skip-gram model using negative sampling.

```
lr, num_epochs = 0.002, 5
train(net, data_iter, lr, num_epochs)
```

```
loss 0.410, 190209.2 tokens/sec on cuda:0
```

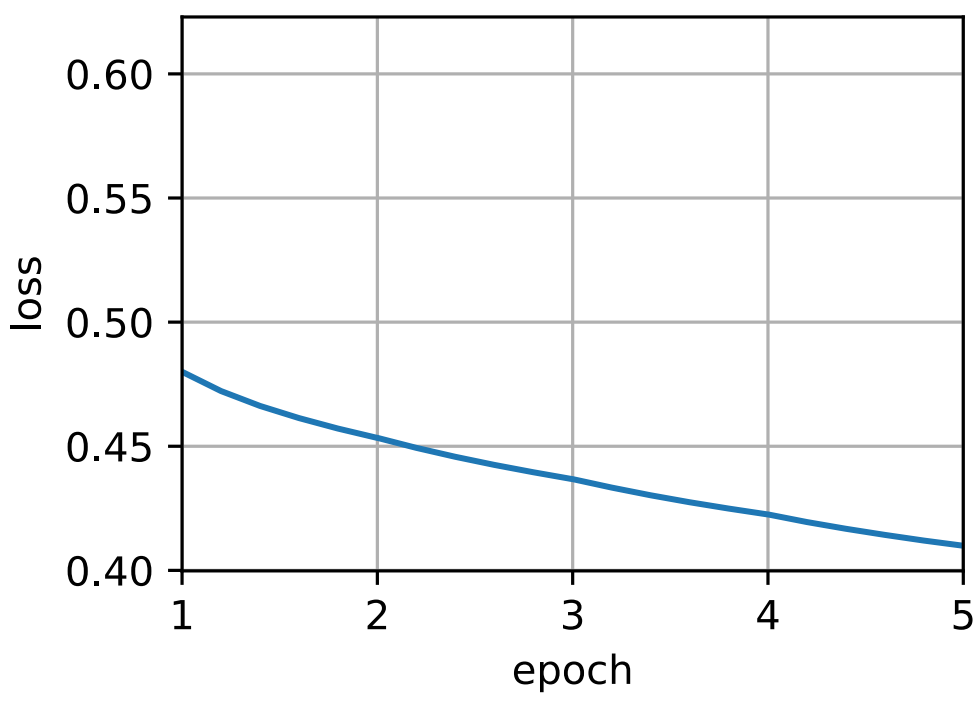

### 15.4.3 Applying Word Embeddings

After training the word2vec model, we can use the cosine similarity of word vectors from the trained model to find words from the dictionary that are most semantically similar to an input word.

```
def get_similar_tokens(query_token, k, embed):
    W = embed.weight.data
    x = W[vocab[query_token]]
    # Compute the cosine similarity. Add 1e-9 for numerical stability
    cos = torch.mv(W, x) / torch.sqrt(torch.sum(W * W, dim=1) *
                                      torch.sum(x * x) + 1e-9)
    topk = torch.topk(cos, k=k+1)[1].cpu().numpy().astype('int32')
    for i in topk[1:]:  # Remove the input words
        print(f'cosine sim={float(cos[i]):.3f}: {vocab.to_tokens(i)}')

get_similar_tokens('chip', 3, net[0])
```

```
cosine sim=0.784: microprocessor
cosine sim=0.743: intel
cosine sim=0.653: chips
```

### 15.4.4 Summary

- We can train a skip-gram model with negative sampling using embedding layers and the binary cross-entropy loss.
- Applications of word embeddings include finding semantically similar words for a given word based on the cosine similarity of word vectors.

### 15.4.5 Exercises

1. Using the trained model, find semantically similar words for other input words. Can you improve the results by tuning hyperparameters?
2. When a training corpus is huge, we often sample context words and noise words for the center words in the current minibatch *when updating model parameters*. In other words, the same center word may have different context words or noise words in different training epochs. What are the benefits of this method? Try to implement this training method.

Discussions[232].

## 15.5 Word Embedding with Global Vectors (GloVe)

Word-word co-occurrences within context windows may carry rich semantic information. For example, in a large corpus word "solid" is more likely to co-occur with "ice" than "steam", but word "gas" probably co-occurs with "steam" more frequently than "ice". Besides, global corpus statistics of such co-occurrences can be precomputed: this can lead to more efficient training. To leverage statistical information in the entire corpus for word embedding, let's first revisit the skip-gram model in Section 15.1.3, but interpreting it using global corpus statistics such as co-occurrence counts.

### 15.5.1 Skip-Gram with Global Corpus Statistics

Denoting by $q_{ij}$ the conditional probability $P(w_j \mid w_i)$ of word $w_j$ given word $w_i$ in the skip-gram model, we have

$$q_{ij} = \frac{\exp(\mathbf{u}_j^\top \mathbf{v}_i)}{\sum_{k \in \mathcal{V}} \exp(\mathbf{u}_k^\top \mathbf{v}_i)}, \tag{15.5.1}$$

where for any index $i$ vectors $\mathbf{v}_i$ and $\mathbf{u}_i$ represent word $w_i$ as the center word and context word, respectively, and $\mathcal{V} = \{0, 1, \ldots, |\mathcal{V}| - 1\}$ is the index set of the vocabulary.

Consider word $w_i$ that may occur multiple times in the corpus. In the entire corpus, all the context words wherever $w_i$ is taken as their center word form a *multiset* $C_i$ of word indices that *allows for multiple instances of the same element*. For any element, its number of instances is called its *multiplicity*. To illustrate with an example, suppose that word $w_i$ occurs twice in the corpus and indices of the context words that take $w_i$ as their center word in the two context windows are $k, j, m, k$ and $k, l, k, j$. Thus, multiset $C_i = \{j, j, k, k, k, k, l, m\}$, where multiplicities of elements $j, k, l, m$ are 2, 4, 1, 1, respectively.

Now let's denote the multiplicity of element $j$ in multiset $C_i$ as $x_{ij}$. This is the global co-occurrence count of word $w_j$ (as the context word) and word $w_i$ (as the center word) in the same context window in the entire corpus. Using such global corpus statistics, the loss function of the skip-gram model is equivalent to

$$-\sum_{i\in\mathcal{V}}\sum_{j\in\mathcal{V}} x_{ij} \log q_{ij}. \tag{15.5.2}$$

We further denote by $x_i$ the number of all the context words in the context windows where $w_i$ occurs as their center word, which is equivalent to $|C_i|$. Letting $p_{ij}$ be the conditional probability $x_{ij}/x_i$ for generating context word $w_j$ given center word $w_i$, (15.5.2) can be rewritten as

$$-\sum_{i\in\mathcal{V}} x_i \sum_{j\in\mathcal{V}} p_{ij} \log q_{ij}. \tag{15.5.3}$$

In (15.5.3), $-\sum_{j\in\mathcal{V}} p_{ij} \log q_{ij}$ calculates the cross-entropy of the conditional distribution $p_{ij}$ of global corpus statistics and the conditional distribution $q_{ij}$ of model predictions. This loss is also weighted by $x_i$ as explained above. Minimizing the loss function in (15.5.3) will allow the predicted conditional distribution to get close to the conditional distribution from the global corpus statistics.

Though being commonly used for measuring the distance between probability distributions, the cross-entropy loss function may not be a good choice here. On the one hand, as we mentioned in Section 15.2, the cost of properly normalizing $q_{ij}$ results in the sum over the entire vocabulary, which can be computationally expensive. On the other hand, a large number of rare events from a large corpus are often modeled by the cross-entropy loss to be assigned with too much weight.

## 15.5.2 The GloVe Model

In view of this, the *GloVe* model makes three changes to the skip-gram model based on squared loss (Pennington *et al.*, 2014):

1. Use variables $p'_{ij} = x_{ij}$ and $q'_{ij} = \exp(\mathbf{u}_j^\top \mathbf{v}_i)$ that are not probability distributions and take the logarithm of both, so the squared loss term is $\left(\log p'_{ij} - \log q'_{ij}\right)^2 = \left(\mathbf{u}_j^\top \mathbf{v}_i - \log x_{ij}\right)^2$.
2. Add two scalar model parameters for each word $w_i$: the center word bias $b_i$ and the context word bias $c_i$.
3. Replace the weight of each loss term with the weight function $h(x_{ij})$, where $h(x)$ is increasing in the interval of $[0, 1]$.

Putting all things together, training GloVe is to minimize the following loss function:

$$\sum_{i\in\mathcal{V}}\sum_{j\in\mathcal{V}} h(x_{ij}) \left(\mathbf{u}_j^\top \mathbf{v}_i + b_i + c_j - \log\, x_{ij}\right)^2. \qquad (15.5.4)$$

For the weight function, a suggested choice is: $h(x) = (x/c)^\alpha$ (e.g $\alpha = 0.75$) if $x < c$ (e.g., $c = 100$); otherwise $h(x) = 1$. In this case, because $h(0) = 0$, the squared loss term for any $x_{ij} = 0$ can be omitted for computational efficiency. For example, when using minibatch stochastic gradient descent for training, at each iteration we randomly sample a minibatch of *non-zero* $x_{ij}$ to calculate gradients and update the model parameters. Note that these non-zero $x_{ij}$ are precomputed global corpus statistics; thus, the model is called GloVe for *Global Vectors*.

It should be emphasized that if word $w_i$ appears in the context window of word $w_j$, then *vice versa*. Therefore, $x_{ij} = x_{ji}$. Unlike word2vec that fits the asymmetric conditional probability $p_{ij}$, GloVe fits the symmetric $\log\, x_{ij}$. Therefore, the center word vector and the context word vector of any word are mathematically equivalent in the GloVe model. However in practice, owing to different initialization values, the same word may still get different values in these two vectors after training: GloVe sums them up as the output vector.

### 15.5.3 Interpreting GloVe from the Ratio of Co-occurrence Probabilities

We can also interpret the GloVe model from another perspective. Using the same notation in Section 15.5.1, let $p_{ij} \stackrel{\text{def}}{=} P(w_j \mid w_i)$ be the conditional probability of generating the context word $w_j$ given $w_i$ as the center word in the corpus. Table 15.5.1 lists several co-occurrence probabilities given words "ice" and "steam" and their ratios based on statistics from a large corpus.

Table 15.5.1: Word-word co-occurrence probabilities and their ratios from a large corpus (adapted from Table 1 in Pennington *et al.* (2014))

| $w_k$= | solid | gas | water | fashion |
|---|---|---|---|---|
| $p_1 = P(w_k \mid \text{ice})$ | 0.00019 | 0.000066 | 0.003 | 0.000017 |
| $p_2 = P(w_k \mid \text{steam})$ | 0.000022 | 0.00078 | 0.0022 | 0.000018 |
| $p_1/p_2$ | 8.9 | 0.085 | 1.36 | 0.96 |

We can observe the following from Table 15.5.1:

- For a word $w_k$ that is related to "ice" but unrelated to "steam", such as $w_k$ = solid, we expect a larger ratio of co-occurence probabilities, such as 8.9.
- For a word $w_k$ that is related to "steam" but unrelated to "ice", such as $w_k$ = gas, we expect a smaller ratio of co-occurence probabilities, such as 0.085.
- For a word $w_k$ that is related to both "ice" and "steam", such as $w_k$ = water, we expect a ratio of co-occurence probabilities that is close to 1, such as 1.36.

- For a word $w_k$ that is unrelated to both "ice" and "steam", such as $w_k =$ fashion, we expect a ratio of co-occurence probabilities that is close to 1, such as 0.96.

It can be seen that the ratio of co-occurrence probabilities can intuitively express the relationship between words. Thus, we can design a function of three word vectors to fit this ratio. For the ratio of co-occurrence probabilities $p_{ij}/p_{ik}$ with $w_i$ being the center word and $w_j$ and $w_k$ being the context words, we want to fit this ratio using some function $f$:

$$f(\mathbf{u}_j, \mathbf{u}_k, \mathbf{v}_i) \approx \frac{p_{ij}}{p_{ik}}. \tag{15.5.5}$$

Among many possible designs for $f$, we only pick a reasonable choice in the following. Since the ratio of co-occurrence probabilities is a scalar, we require that $f$ be a scalar function, such as $f(\mathbf{u}_j, \mathbf{u}_k, \mathbf{v}_i) = f\left((\mathbf{u}_j - \mathbf{u}_k)^\top \mathbf{v}_i\right)$. Switching word indices $j$ and $k$ in (15.5.5), it must hold that $f(x)f(-x) = 1$, so one possibility is $f(x) = \exp(x)$, i.e.,

$$f(\mathbf{u}_j, \mathbf{u}_k, \mathbf{v}_i) = \frac{\exp\left(\mathbf{u}_j^\top \mathbf{v}_i\right)}{\exp\left(\mathbf{u}_k^\top \mathbf{v}_i\right)} \approx \frac{p_{ij}}{p_{ik}}. \tag{15.5.6}$$

Now let's pick $\exp\left(\mathbf{u}_j^\top \mathbf{v}_i\right) \approx \alpha p_{ij}$, where $\alpha$ is a constant. Since $p_{ij} = x_{ij}/x_i$, after taking the logarithm on both sides we get $\mathbf{u}_j^\top \mathbf{v}_i \approx \log\, \alpha + \log\, x_{ij} - \log\, x_i$. We may use additional bias terms to fit $-\log\, \alpha + \log\, x_i$, such as the center word bias $b_i$ and the context word bias $c_j$:

$$\mathbf{u}_j^\top \mathbf{v}_i + b_i + c_j \approx \log\, x_{ij}. \tag{15.5.7}$$

Measuring the squared error of (15.5.7) with weights, the GloVe loss function in (15.5.4) is obtained.

## 15.5.4 Summary

- The skip-gram model can be interpreted using global corpus statistics such as word-word co-occurrence counts.
- The cross-entropy loss may not be a good choice for measuring the difference of two probability distributions, especially for a large corpus. GloVe uses squared loss to fit precomputed global corpus statistics.
- The center word vector and the context word vector are mathematically equivalent for any word in GloVe.
- GloVe can be interpreted from the ratio of word-word co-occurrence probabilities.

## 15.5.5 Exercises

1. If words $w_i$ and $w_j$ co-occur in the same context window, how can we use their distance in the text sequence to redesign the method for calculating the conditional probability $p_{ij}$? Hint: see Section 4.2 of the GloVe paper (Pennington *et al.*, 2014).

2. For any word, are its center word bias and context word bias mathematically equivalent in GloVe? Why?

Discussions[233].

# 15.6 Subword Embedding

In English, words such as “helps”, “helped”, and “helping” are inflected forms of the same word “help”. The relationship between “dog” and “dogs” is the same as that between “cat” and “cats”, and the relationship between “boy” and “boyfriend” is the same as that between “girl” and “girlfriend”. In other languages such as French and Spanish, many verbs have over 40 inflected forms, while in Finnish, a noun may have up to 15 cases. In linguistics, morphology studies word formation and word relationships. However, the internal structure of words was neither explored in word2vec nor in GloVe.

## 15.6.1 The fastText Model

Recall how words are represented in word2vec. In both the skip-gram model and the continuous bag-of-words model, different inflected forms of the same word are directly represented by different vectors without shared parameters. To use morphological information, the *fastText* model proposed a *subword embedding* approach, where a subword is a character $n$-gram (Bojanowski *et al.*, 2017). Instead of learning word-level vector representations, fastText can be considered as the subword-level skip-gram, where each *center word* is represented by the sum of its subword vectors.

Let’s illustrate how to obtain subwords for each center word in fastText using the word “where”. First, add special characters “<” and “>” at the beginning and end of the word to distinguish prefixes and suffixes from other subwords. Then, extract character $n$-grams from the word. For example, when $n = 3$, we obtain all subwords of length 3: “<wh”, “whe”, “her”, “ere”, “re>”, and the special subword “<where>”.

In fastText, for any word $w$, denote by $\mathcal{G}_w$ the union of all its subwords of length between 3 and 6 and its special subword. The vocabulary is the union of the subwords of all words. Letting $\mathbf{z}_g$ be the vector of subword $g$ in the dictionary, the vector $\mathbf{v}_w$ for word $w$ as a center word in the skip-gram model is the sum of its subword vectors:

$$\mathbf{v}_w = \sum_{g \in \mathcal{G}_w} \mathbf{z}_g. \tag{15.6.1}$$

The rest of fastText is the same as the skip-gram model. Compared with the skip-gram model, the vocabulary in fastText is larger, resulting in more model parameters. Besides, to calculate the representation of a word, all its subword vectors have to be summed, leading to higher computational complexity. However, thanks to shared parameters from subwords among words with similar structures, rare words and even out-of-vocabulary words may obtain better vector representations in fastText.

## 15.6.2 Byte Pair Encoding

In fastText, all the extracted subwords have to be of the specified lengths, such as 3 to 6, thus the vocabulary size cannot be predefined. To allow for variable-length subwords in a fixed-size vocabulary, we can apply a compression algorithm called *byte pair encoding* (BPE) to extract subwords (Sennrich *et al.*, 2015).

Byte pair encoding performs a statistical analysis of the training dataset to discover common symbols within a word, such as consecutive characters of arbitrary length. Starting from symbols of length 1, byte pair encoding iteratively merges the most frequent pair of consecutive symbols to produce new longer symbols. Note that for efficiency, pairs crossing word boundaries are not considered. In the end, we can use such symbols as subwords to segment words. Byte pair encoding and its variants has been used for input representations in popular natural language processing pretraining models such as GPT-2 (Radford *et al.*, 2019) and RoBERTa (Liu *et al.*, 2019). In the following, we will illustrate how byte pair encoding works.

First, we initialize the vocabulary of symbols as all the English lowercase characters, a special end-of-word symbol `'_'`, and a special unknown symbol `'[UNK]'`.

```
import collections

symbols = ['a', 'b', 'c', 'd', 'e', 'f', 'g', 'h', 'i', 'j', 'k', 'l', 'm',
           'n', 'o', 'p', 'q', 'r', 's', 't', 'u', 'v', 'w', 'x', 'y', 'z',
           '_', '[UNK]']
```

Since we do not consider symbol pairs that cross boundaries of words, we only need a dictionary `raw_token_freqs` that maps words to their frequencies (number of occurrences) in a dataset. Note that the special symbol `'_'` is appended to each word so that we can easily recover a word sequence (e.g., "a taller man") from a sequence of output symbols ( e.g., "a_ tall er_ man"). Since we start the merging process from a vocabulary of only single characters and special symbols, space is inserted between every pair of consecutive characters within each word (keys of the dictionary `token_freqs`). In other words, space is the delimiter between symbols within a word.

```
raw_token_freqs = {'fast_': 4, 'faster_': 3, 'tall_': 5, 'taller_': 4}
token_freqs = {}
for token, freq in raw_token_freqs.items():
    token_freqs[' '.join(list(token))] = raw_token_freqs[token]
token_freqs
```

```
{'f a s t _': 4, 'f a s t e r _': 3, 't a l l _': 5, 't a l l e r _': 4}
```

We define the following `get_max_freq_pair` function that returns the most frequent pair of consecutive symbols within a word, where words come from keys of the input dictionary `token_freqs`.

```
def get_max_freq_pair(token_freqs):
    pairs = collections.defaultdict(int)
    for token, freq in token_freqs.items():
        symbols = token.split()
        for i in range(len(symbols) - 1):
            # Key of `pairs` is a tuple of two consecutive symbols
            pairs[symbols[i], symbols[i + 1]] += freq
    return max(pairs, key=pairs.get)  # Key of `pairs` with the max value
```

As a greedy approach based on frequency of consecutive symbols, byte pair encoding will use the following `merge_symbols` function to merge the most frequent pair of consecutive symbols to produce new symbols.

```
def merge_symbols(max_freq_pair, token_freqs, symbols):
    symbols.append(''.join(max_freq_pair))
    new_token_freqs = dict()
    for token, freq in token_freqs.items():
        new_token = token.replace(' '.join(max_freq_pair),
                                  ''.join(max_freq_pair))
        new_token_freqs[new_token] = token_freqs[token]
    return new_token_freqs
```

Now we iteratively perform the byte pair encoding algorithm over the keys of the dictionary `token_freqs`. In the first iteration, the most frequent pair of consecutive symbols are `'t'` and `'a'`, thus byte pair encoding merges them to produce a new symbol `'ta'`. In the second iteration, byte pair encoding continues to merge `'ta'` and `'l'` to result in another new symbol `'tal'`.

```
num_merges = 10
for i in range(num_merges):
    max_freq_pair = get_max_freq_pair(token_freqs)
    token_freqs = merge_symbols(max_freq_pair, token_freqs, symbols)
    print(f'merge #{i + 1}:', max_freq_pair)
```

```
merge #1: ('t', 'a')
merge #2: ('ta', 'l')
merge #3: ('tal', 'l')
merge #4: ('f', 'a')
merge #5: ('fa', 's')
merge #6: ('fas', 't')
merge #7: ('e', 'r')
merge #8: ('er', '_')
merge #9: ('tall', '_')
merge #10: ('fast', '_')
```

After 10 iterations of byte pair encoding, we can see that list `symbols` now contains 10 more symbols that are iteratively merged from other symbols.

```
print(symbols)
```

```
['a', 'b', 'c', 'd', 'e', 'f', 'g', 'h', 'i', 'j', 'k', 'l', 'm', 'n', 'o', 'p
↪', 'q', 'r', 's', 't', 'u', 'v', 'w', 'x', 'y', 'z', '_', '[UNK]', 'ta', 'tal
↪', 'tall', 'fa', 'fas', 'fast', 'er', 'er_', 'tall_', 'fast_']
```

For the same dataset specified in the keys of the dictionary `raw_token_freqs`, each word in the dataset is now segmented by subwords “fast_”, “fast”, “er_”, “tall_”, and “tall” as a result of the byte pair encoding algorithm. For instance, words “faster_” and “taller_” are segmented as “fast er_” and “tall er_”, respectively.

```
print(list(token_freqs.keys()))
```

```
['fast_', 'fast er_', 'tall_', 'tall er_']
```

Note that the result of byte pair encoding depends on the dataset being used. We can also use the subwords learned from one dataset to segment words of another dataset. As a greedy approach, the following `segment_BPE` function tries to break words into the longest possible subwords from the input argument `symbols`.

```
def segment_BPE(tokens, symbols):
    outputs = []
    for token in tokens:
        start, end = 0, len(token)
        cur_output = []
        # Segment token with the longest possible subwords from symbols
        while start < len(token) and start < end:
            if token[start: end] in symbols:
                cur_output.append(token[start: end])
                start = end
                end = len(token)
            else:
                end -= 1
        if start < len(token):
            cur_output.append('[UNK]')
        outputs.append(' '.join(cur_output))
    return outputs
```

In the following, we use the subwords in list `symbols`, which is learned from the aforementioned dataset, to segment `tokens` that represent another dataset.

```
tokens = ['tallest_', 'fatter_']
print(segment_BPE(tokens, symbols))
```

```
['tall e s t _', 'fa t t er_']
```

### 15.6.3 Summary

- The fastText model proposes a subword embedding approach. Based on the skip-gram model in word2vec, it represents a center word as the sum of its subword vectors.
- Byte pair encoding performs a statistical analysis of the training dataset to discover common symbols within a word. As a greedy approach, byte pair encoding iteratively merges the most frequent pair of consecutive symbols.
- Subword embedding may improve the quality of representations of rare words and out-of-dictionary words.

### 15.6.4 Exercises

1. As an example, there are about $3 \times 10^8$ possible 6-grams in English. What is the issue when there are too many subwords? How to address the issue? Hint: refer to the end of Section 3.2 of the fastText paper (Bojanowski *et al.*, 2017).
2. How to design a subword embedding model based on the continuous bag-of-words model?
3. To get a vocabulary of size $m$, how many merging operations are needed when the initial symbol vocabulary size is $n$?
4. How to extend the idea of byte pair encoding to extract phrases?

Discussions[234].

## 15.7 Word Similarity and Analogy

In Section 15.4, we trained a word2vec model on a small dataset, and applied it to find semantically similar words for an input word. In practice, word vectors that are pretrained on large corpora can be applied to downstream natural language processing tasks, which will be covered later in Chapter 16. To demonstrate semantics of pretrained word vectors from large corpora in a straightforward way, let's apply them in the word similarity and analogy tasks.

```
import os
import torch
from torch import nn
from d2l import torch as d2l
```

### 15.7.1 Loading Pretrained Word Vectors

Below lists pretrained GloVe embeddings of dimension 50, 100, and 300, which can be downloaded from the GloVe website[235]. The pretrained fastText embeddings are available in multiple languages. Here we consider one English version (300-dimensional "wiki.en") that can be downloaded from the fastText website[236].

235

236

```
#@save
d2l.DATA_HUB['glove.6b.50d'] = (d2l.DATA_URL + 'glove.6B.50d.zip',
                                '0b8703943ccdb6eb788e6f091b8946e82231bc4d')

#@save
d2l.DATA_HUB['glove.6b.100d'] = (d2l.DATA_URL + 'glove.6B.100d.zip',
                                 'cd43bfb07e44e6f27cbcc7bc9ae3d80284fdaf5a')

#@save
d2l.DATA_HUB['glove.42b.300d'] = (d2l.DATA_URL + 'glove.42B.300d.zip',
                                  'b5116e234e9eb9076672cfeabf5469f3eec904fa')

#@save
d2l.DATA_HUB['wiki.en'] = (d2l.DATA_URL + 'wiki.en.zip',
                           'c1816da3821ae9f43899be655002f6c723e91b88')
```

To load these pretrained GloVe and fastText embeddings, we define the following `TokenEmbedding` class.

```
#@save
class TokenEmbedding:
    """Token Embedding."""
    def __init__(self, embedding_name):
        self.idx_to_token, self.idx_to_vec = self._load_embedding(
            embedding_name)
        self.unknown_idx = 0
        self.token_to_idx = {token: idx for idx, token in
                             enumerate(self.idx_to_token)}

    def _load_embedding(self, embedding_name):
        idx_to_token, idx_to_vec = ['<unk>'], []
        data_dir = d2l.download_extract(embedding_name)
        # GloVe website: https://nlp.stanford.edu/projects/glove/
        # fastText website: https://fasttext.cc/
        with open(os.path.join(data_dir, 'vec.txt'), 'r') as f:
            for line in f:
                elems = line.rstrip().split(' ')
                token, elems = elems[0], [float(elem) for elem in elems[1:]]
                # Skip header information, such as the top row in fastText
                if len(elems) > 1:
                    idx_to_token.append(token)
                    idx_to_vec.append(elems)
        idx_to_vec = [[0] * len(idx_to_vec[0])] + idx_to_vec
        return idx_to_token, torch.tensor(idx_to_vec)

    def __getitem__(self, tokens):
```

```
        indices = [self.token_to_idx.get(token, self.unknown_idx)
                   for token in tokens]
        vecs = self.idx_to_vec[torch.tensor(indices)]
        return vecs

    def __len__(self):
        return len(self.idx_to_token)
```

Below we load the 50-dimensional GloVe embeddings (pretrained on a Wikipedia subset). When creating the `TokenEmbedding` instance, the specified embedding file has to be downloaded if it was not yet.

```
glove_6b50d = TokenEmbedding('glove.6b.50d')
```

Output the vocabulary size. The vocabulary contains 400000 words (tokens) and a special unknown token.

```
len(glove_6b50d)
```

```
400001
```

We can get the index of a word in the vocabulary, and vice versa.

```
glove_6b50d.token_to_idx['beautiful'], glove_6b50d.idx_to_token[3367]
```

```
(3367, 'beautiful')
```

## 15.7.2 Applying Pretrained Word Vectors

Using the loaded GloVe vectors, we will demonstrate their semantics by applying them in the following word similarity and analogy tasks.

### Word Similarity

Similar to Section 15.4.3, in order to find semantically similar words for an input word based on cosine similarities between word vectors, we implement the following `knn` ($k$-nearest neighbors) function.

```
def knn(W, x, k):
    # Add 1e-9 for numerical stability
    cos = torch.mv(W, x.reshape(-1,)) / (
```

```
        torch.sqrt(torch.sum(W * W, axis=1) + 1e-9) *
        torch.sqrt((x * x).sum()))
    _, topk = torch.topk(cos, k=k)
    return topk, [cos[int(i)] for i in topk]
```

Then, we search for similar words using the pretrained word vectors from the `TokenEmbedding` instance `embed`.

```
def get_similar_tokens(query_token, k, embed):
    topk, cos = knn(embed.idx_to_vec, embed[[query_token]], k + 1)
    for i, c in zip(topk[1:], cos[1:]):  # Exclude the input word
        print(f'cosine sim={float(c):.3f}: {embed.idx_to_token[int(i)]}')
```

The vocabulary of the pretrained word vectors in `glove_6b50d` contains 400000 words and a special unknown token. Excluding the input word and unknown token, among this vocabulary let's find three most semantically similar words to word "chip".

```
get_similar_tokens('chip', 3, glove_6b50d)
```

```
cosine sim=0.856: chips
cosine sim=0.749: intel
cosine sim=0.749: electronics
```

Below outputs similar words to "baby" and "beautiful".

```
get_similar_tokens('baby', 3, glove_6b50d)
```

```
cosine sim=0.839: babies
cosine sim=0.800: boy
cosine sim=0.792: girl
```

```
get_similar_tokens('beautiful', 3, glove_6b50d)
```

```
cosine sim=0.921: lovely
cosine sim=0.893: gorgeous
cosine sim=0.830: wonderful
```

## Word Analogy

Besides finding similar words, we can also apply word vectors to word analogy tasks. For example, "man":"woman"::"son":"daughter" is the form of a word analogy: "man" is to "woman" as "son" is to "daughter". Specifically, the word analogy completion task can be defined as: for a word analogy $a : b :: c : d$, given the first three words $a$, $b$ and $c$, find $d$. Denote the vector of word $w$ by $\text{vec}(w)$. To complete the analogy, we will find the word whose vector is most similar to the result of $\text{vec}(c) + \text{vec}(b) - \text{vec}(a)$.

```
def get_analogy(token_a, token_b, token_c, embed):
    vecs = embed[[token_a, token_b, token_c]]
    x = vecs[1] - vecs[0] + vecs[2]
    topk, cos = knn(embed.idx_to_vec, x, 1)
    return embed.idx_to_token[int(topk[0])]  # Remove unknown words
```

Let's verify the "male-female" analogy using the loaded word vectors.

```
get_analogy('man', 'woman', 'son', glove_6b50d)
```

```
'daughter'
```

Below completes a "capital-country" analogy: "beijing":"china"::"tokyo":"japan". This demonstrates semantics in the pretrained word vectors.

```
get_analogy('beijing', 'china', 'tokyo', glove_6b50d)
```

```
'japan'
```

For the "adjective-superlative adjective" analogy such as "bad":"worst"::"big":"biggest", we can see that the pretrained word vectors may capture the syntactic information.

```
get_analogy('bad', 'worst', 'big', glove_6b50d)
```

```
'biggest'
```

To show the captured notion of past tense in the pretrained word vectors, we can test the syntax using the "present tense-past tense" analogy: "do":"did"::"go":"went".

```
get_analogy('do', 'did', 'go', glove_6b50d)
```

```
'went'
```

## 15.7.3 Summary

- In practice, word vectors that are pretrained on large corpora can be applied to downstream natural language processing tasks.
- Pretrained word vectors can be applied to the word similarity and analogy tasks.

### 15.7.4 Exercises

1. Test the fastText results using `TokenEmbedding('wiki.en')`.
2. When the vocabulary is extremely large, how can we find similar words or complete a word analogy faster?

Discussions[237].

## 15.8 Bidirectional Encoder Representations from Transformers (BERT)

We have introduced several word embedding models for natural language understanding. After pretraining, the output can be thought of as a matrix where each row is a vector that represents a word of a predefined vocabulary. In fact, these word embedding models are all *context-independent*. Let's begin by illustrating this property.

### 15.8.1 From Context-Independent to Context-Sensitive

Recall the experiments in Section 15.4 and Section 15.7. For instance, word2vec and GloVe both assign the same pretrained vector to the same word regardless of the context of the word (if any). Formally, a context-independent representation of any token $x$ is a function $f(x)$ that only takes $x$ as its input. Given the abundance of polysemy and complex semantics in natural languages, context-independent representations have obvious limitations. For instance, the word "crane" in contexts "a crane is flying" and "a crane driver came" has completely different meanings; thus, the same word may be assigned different representations depending on contexts.

This motivates the development of *context-sensitive* word representations, where representations of words depend on their contexts. Hence, a context-sensitive representation of token $x$ is a function $f(x, c(x))$ depending on both $x$ and its context $c(x)$. Popular context-sensitive representations include TagLM (language-model-augmented sequence tagger) (Peters *et al.*, 2017), CoVe (Context Vectors) (McCann *et al.*, 2017), and ELMo (Embeddings from Language Models) (Peters *et al.*, 2018).

For example, by taking the entire sequence as input, ELMo is a function that assigns a representation to each word from the input sequence. Specifically, ELMo combines all the intermediate layer representations from pretrained bidirectional LSTM as the output representation. Then the ELMo representation will be added to a downstream task's existing supervised model as additional features, such as by concatenating ELMo representation and the original representation (e.g., GloVe) of tokens in the existing model. On the one hand, all the weights in the pretrained bidirectional LSTM model are frozen after ELMo representations are added. On the other hand, the existing supervised model is specifically customized for a given task. Leveraging different best models for different tasks at that time, adding ELMo improved the state of the art across six natural language processing tasks: sentiment analysis, natural language inference, semantic role labeling, coreference resolution, named entity recognition, and question answering.

### 15.8.2 From Task-Specific to Task-Agnostic

Although ELMo has significantly improved solutions to a diverse set of natural language processing tasks, each solution still hinges on a *task-specific* architecture. However, it is practically non-trivial to craft a specific architecture for every natural language processing task. The GPT (Generative Pre-Training) model represents an effort in designing a general *task-agnostic* model for context-sensitive representations (Radford *et al.*, 2018). Built on a Transformer decoder, GPT pretrains a language model that will be used to represent text sequences. When applying GPT to a downstream task, the output of the language model will be fed into an added linear output layer to predict the label of the task. In sharp contrast to ELMo that freezes parameters of the pretrained model, GPT fine-tunes *all* the parameters in the pretrained Transformer decoder during supervised learning of the downstream task. GPT was evaluated on twelve tasks of natural language inference, question answering, sentence similarity, and classification, and improved the state of the art in nine of them with minimal changes to the model architecture.

However, due to the autoregressive nature of language models, GPT only looks forward (left-to-right). In contexts "i went to the bank to deposit cash" and "i went to the bank to sit down", as "bank" is sensitive to the context to its left, GPT will return the same representation for "bank", though it has different meanings.

### 15.8.3 BERT: Combining the Best of Both Worlds

As we have seen, ELMo encodes context bidirectionally but uses task-specific architectures; while GPT is task-agnostic but encodes context left-to-right. Combining the best of both worlds, BERT (Bidirectional Encoder Representations from Transformers) encodes context bidirectionally and requires minimal architecture changes for a wide range of natural language processing tasks (Devlin *et al.*, 2018). Using a pretrained Transformer encoder, BERT is able to represent any token based on its bidirectional context. During supervised learning of downstream tasks, BERT is similar to GPT in two aspects. First, BERT representations will be fed into an added output layer, with minimal changes to the model architecture depending on nature of tasks, such as predicting for every token vs. predicting for the entire sequence. Second, all the parameters of the pretrained Transformer encoder are fine-tuned, while the additional output layer will be trained from scratch. Fig. 15.8.1 depicts the differences among ELMo, GPT, and BERT.

BERT further improved the state of the art on eleven natural language processing tasks under broad categories of (i) single text classification (e.g., sentiment analysis), (ii) text pair classification (e.g., natural language inference), (iii) question answering, (iv) text tagging (e.g., named entity recognition). All proposed in 2018, from context-sensitive ELMo to task-agnostic GPT and BERT, conceptually simple yet empirically powerful pretraining of deep representations for natural languages have revolutionized solutions to various natural language processing tasks.

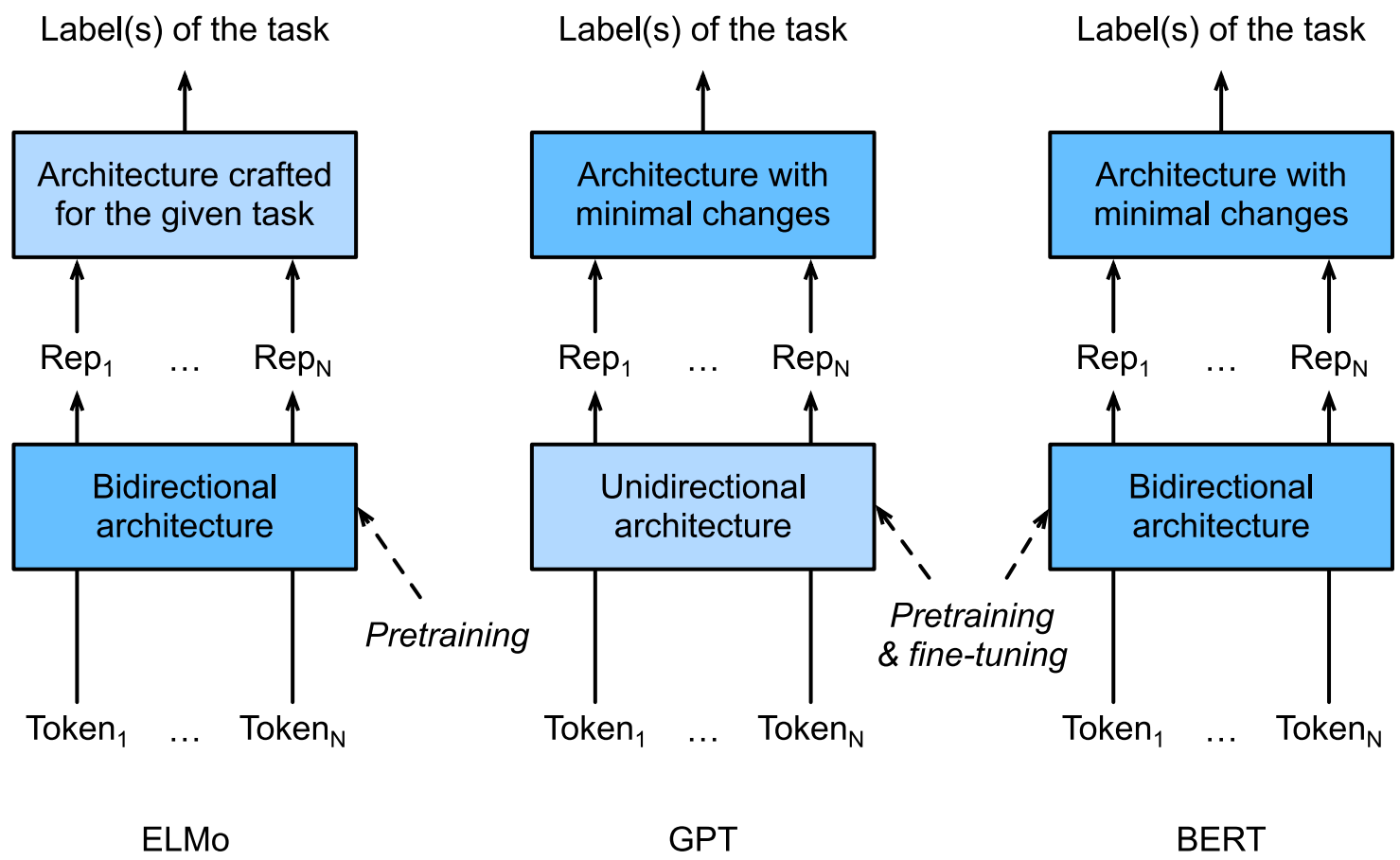

Fig. 15.8.1 A comparison of ELMo, GPT, and BERT.

In the rest of this chapter, we will dive into the pretraining of BERT. When natural language processing applications are explained in Chapter 16, we will illustrate fine-tuning of BERT for downstream applications.

```
import torch
from torch import nn
from d2l import torch as d2l
```

## 15.8.4 Input Representation

In natural language processing, some tasks (e.g., sentiment analysis) take single text as input, while in some other tasks (e.g., natural language inference), the input is a pair of text sequences. The BERT input sequence unambiguously represents both single text and text pairs. In the former, the BERT input sequence is the concatenation of the special classification token "<cls>", tokens of a text sequence, and the special separation token "<sep>". In the latter, the BERT input sequence is the concatenation of "<cls>", tokens of the first text sequence, "<sep>", tokens of the second text sequence, and "<sep>". We will consistently distinguish the terminology "BERT input sequence" from other types of "sequences". For instance, one *BERT input sequence* may include either one *text sequence* or two *text sequences*.

To distinguish text pairs, the learned segment embeddings $\mathbf{e}_A$ and $\mathbf{e}_B$ are added to the token embeddings of the first sequence and the second sequence, respectively. For single text inputs, only $\mathbf{e}_A$ is used.

The following `get_tokens_and_segments` takes either one sentence or two sentences as input, then returns tokens of the BERT input sequence and their corresponding segment IDs.

```
#@save
def get_tokens_and_segments(tokens_a, tokens_b=None):
    """Get tokens of the BERT input sequence and their segment IDs."""
    tokens = ['<cls>'] + tokens_a + ['<sep>']
    # 0 and 1 are marking segment A and B, respectively
    segments = [0] * (len(tokens_a) + 2)
    if tokens_b is not None:
        tokens += tokens_b + ['<sep>']
        segments += [1] * (len(tokens_b) + 1)
    return tokens, segments
```

BERT chooses the Transformer encoder as its bidirectional architecture. Common in the Transformer encoder, positional embeddings are added at every position of the BERT input sequence. However, different from the original Transformer encoder, BERT uses *learnable* positional embeddings. To sum up, Fig. 15.8.2 shows that the embeddings of the BERT input sequence are the sum of the token embeddings, segment embeddings, and positional embeddings.

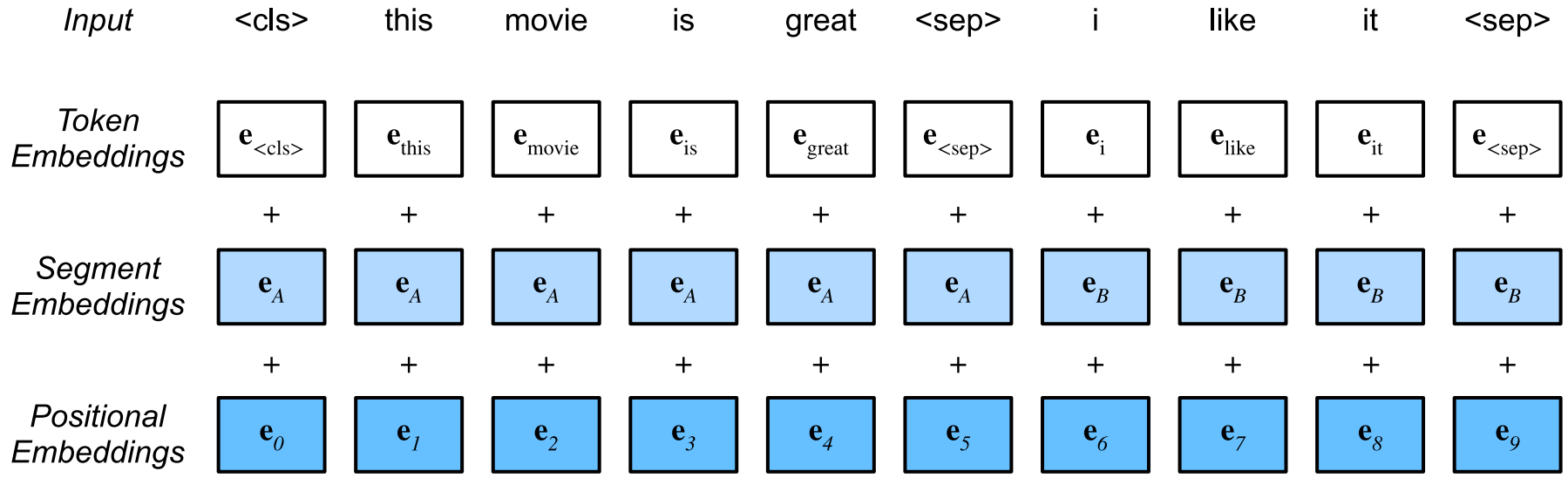

Fig. 15.8.2 The embeddings of the BERT input sequence are the sum of the token embeddings, segment embeddings, and positional embeddings.

The following `BERTEncoder` class is similar to the `TransformerEncoder` class as implemented in Section 11.7. Different from `TransformerEncoder`, `BERTEncoder` uses segment embeddings and learnable positional embeddings.

```
#@save
class BERTEncoder(nn.Module):
    """BERT encoder."""
    def __init__(self, vocab_size, num_hiddens, ffn_num_hiddens, num_heads,
                 num_blks, dropout, max_len=1000, **kwargs):
        super(BERTEncoder, self).__init__(**kwargs)
        self.token_embedding = nn.Embedding(vocab_size, num_hiddens)
        self.segment_embedding = nn.Embedding(2, num_hiddens)
        self.blks = nn.Sequential()
        for i in range(num_blks):
            self.blks.add_module(f"{i}", d2l.TransformerEncoderBlock(
```

```
                num_hiddens, ffn_num_hiddens, num_heads, dropout, True))
        # In BERT, positional embeddings are learnable, thus we create a
        # parameter of positional embeddings that are long enough
        self.pos_embedding = nn.Parameter(torch.randn(1, max_len,
                                                      num_hiddens))

    def forward(self, tokens, segments, valid_lens):
        # Shape of `X` remains unchanged in the following code snippet:
        # (batch size, max sequence length, `num_hiddens`)
        X = self.token_embedding(tokens) + self.segment_embedding(segments)
        X = X + self.pos_embedding[:, :X.shape[1], :]
        for blk in self.blks:
            X = blk(X, valid_lens)
        return X
```

Suppose that the vocabulary size is 10000. To demonstrate forward inference of `BERTEncoder`, let's create an instance of it and initialize its parameters.

```
vocab_size, num_hiddens, ffn_num_hiddens, num_heads = 10000, 768, 1024, 4
ffn_num_input, num_blks, dropout = 768, 2, 0.2
encoder = BERTEncoder(vocab_size, num_hiddens, ffn_num_hiddens, num_heads,
                      num_blks, dropout)
```

We define `tokens` to be 2 BERT input sequences of length 8, where each token is an index of the vocabulary. The forward inference of `BERTEncoder` with the input `tokens` returns the encoded result where each token is represented by a vector whose length is predefined by the hyperparameter `num_hiddens`. This hyperparameter is usually referred to as the *hidden size* (number of hidden units) of the Transformer encoder.

```
tokens = torch.randint(0, vocab_size, (2, 8))
segments = torch.tensor([[0, 0, 0, 0, 1, 1, 1, 1], [0, 0, 0, 1, 1, 1, 1, 1]])
encoded_X = encoder(tokens, segments, None)
encoded_X.shape
```

```
torch.Size([2, 8, 768])
```

## 15.8.5 Pretraining Tasks

The forward inference of `BERTEncoder` gives the BERT representation of each token of the input text and the inserted special tokens "<cls>" and "<seq>". Next, we will use these representations to compute the loss function for pretraining BERT. The pretraining is composed of the following two tasks: masked language modeling and next sentence prediction.

### Masked Language Modeling

As illustrated in Section 9.3, a language model predicts a token using the context on its left. To encode context bidirectionally for representing each token, BERT randomly masks tokens and uses tokens from the bidirectional context to predict the masked tokens in a self-supervised fashion. This task is referred to as a *masked language model*.

In this pretraining task, 15% of tokens will be selected at random as the masked tokens for prediction. To predict a masked token without cheating by using the label, one straightforward approach is to always replace it with a special “<mask>” token in the BERT input sequence. However, the artificial special token “<mask>” will never appear in fine-tuning. To avoid such a mismatch between pretraining and fine-tuning, if a token is masked for prediction (e.g., “great” is selected to be masked and predicted in “this movie is great”), in the input it will be replaced with:

- a special “<mask>” token for 80% of the time (e.g., “this movie is great” becomes “this movie is <mask>”);
- a random token for 10% of the time (e.g., “this movie is great” becomes “this movie is drink”);
- the unchanged label token for 10% of the time (e.g., “this movie is great” becomes “this movie is great”).

Note that for 10% of 15% time a random token is inserted. This occasional noise encourages BERT to be less biased towards the masked token (especially when the label token remains unchanged) in its bidirectional context encoding.

We implement the following `MaskLM` class to predict masked tokens in the masked language model task of BERT pretraining. The prediction uses a one-hidden-layer MLP (`self.mlp`). In forward inference, it takes two inputs: the encoded result of `BERTEncoder` and the token positions for prediction. The output is the prediction results at these positions.

```python
#@save
class MaskLM(nn.Module):
    """The masked language model task of BERT."""
    def __init__(self, vocab_size, num_hiddens, **kwargs):
        super(MaskLM, self).__init__(**kwargs)
        self.mlp = nn.Sequential(nn.LazyLinear(num_hiddens),
                                 nn.ReLU(),
                                 nn.LayerNorm(num_hiddens),
                                 nn.LazyLinear(vocab_size))

    def forward(self, X, pred_positions):
        num_pred_positions = pred_positions.shape[1]
        pred_positions = pred_positions.reshape(-1)
        batch_size = X.shape[0]
        batch_idx = torch.arange(0, batch_size)
        # Suppose that `batch_size` = 2, `num_pred_positions` = 3, then
        # `batch_idx` is `torch.tensor([0, 0, 0, 1, 1, 1])`
        batch_idx = torch.repeat_interleave(batch_idx, num_pred_positions)
```

```
        masked_X = X[batch_idx, pred_positions]
        masked_X = masked_X.reshape((batch_size, num_pred_positions, -1))
        mlm_Y_hat = self.mlp(masked_X)
        return mlm_Y_hat
```

To demonstrate the forward inference of `MaskLM`, we create its instance `mlm` and initialize it. Recall that `encoded_X` from the forward inference of `BERTEncoder` represents 2 BERT input sequences. We define `mlm_positions` as the 3 indices to predict in either BERT input sequence of `encoded_X`. The forward inference of `mlm` returns prediction results `mlm_Y_hat` at all the masked positions `mlm_positions` of `encoded_X`. For each prediction, the size of the result is equal to the vocabulary size.

```
mlm = MaskLM(vocab_size, num_hiddens)
mlm_positions = torch.tensor([[1, 5, 2], [6, 1, 5]])
mlm_Y_hat = mlm(encoded_X, mlm_positions)
mlm_Y_hat.shape
```

```
torch.Size([2, 3, 10000])
```

With the ground truth labels `mlm_Y` of the predicted tokens `mlm_Y_hat` under masks, we can calculate the cross-entropy loss of the masked language model task in BERT pretraining.

```
mlm_Y = torch.tensor([[7, 8, 9], [10, 20, 30]])
loss = nn.CrossEntropyLoss(reduction='none')
mlm_l = loss(mlm_Y_hat.reshape((-1, vocab_size)), mlm_Y.reshape(-1))
mlm_l.shape
```

```
torch.Size([6])
```

## Next Sentence Prediction

Although masked language modeling is able to encode bidirectional context for representing words, it does not explicitly model the logical relationship between text pairs. To help understand the relationship between two text sequences, BERT considers a binary classification task, *next sentence prediction*, in its pretraining. When generating sentence pairs for pretraining, for half of the time they are indeed consecutive sentences with the label "True"; while for the other half of the time the second sentence is randomly sampled from the corpus with the label "False".

The following `NextSentencePred` class uses a one-hidden-layer MLP to predict whether the second sentence is the next sentence of the first in the BERT input sequence. Due to self-attention in the Transformer encoder, the BERT representation of the special token "<cls>" encodes both the two sentences from the input. Hence, the output layer (`self.output`) of the MLP classifier takes `X` as input, where `X` is the output of the MLP hidden layer whose input is the encoded “<cls>” token.

```
#@save
class NextSentencePred(nn.Module):
    """The next sentence prediction task of BERT."""
    def __init__(self, **kwargs):
        super(NextSentencePred, self).__init__(**kwargs)
        self.output = nn.LazyLinear(2)

    def forward(self, X):
        # `X` shape: (batch size, `num_hiddens`)
        return self.output(X)
```

We can see that the forward inference of an `NextSentencePred` instance returns binary predictions for each BERT input sequence.

```
# PyTorch by default will not flatten the tensor as seen in mxnet where, if
# flatten=True, all but the first axis of input data are collapsed together
encoded_X = torch.flatten(encoded_X, start_dim=1)
# input_shape for NSP: (batch size, `num_hiddens`)
nsp = NextSentencePred()
nsp_Y_hat = nsp(encoded_X)
nsp_Y_hat.shape
```

```
torch.Size([2, 2])
```

The cross-entropy loss of the 2 binary classifications can also be computed.

```
nsp_y = torch.tensor([0, 1])
nsp_l = loss(nsp_Y_hat, nsp_y)
nsp_l.shape
```

```
torch.Size([2])
```

It is noteworthy that all the labels in both the aforementioned pretraining tasks can be trivially obtained from the pretraining corpus without manual labeling effort. The original BERT has been pretrained on the concatenation of BookCorpus (Zhu *et al.*, 2015) and English Wikipedia. These two text corpora are huge: they have 800 million words and 2.5 billion words, respectively.

### 15.8.6 Putting It All Together

When pretraining BERT, the final loss function is a linear combination of both the loss functions for masked language modeling and next sentence prediction. Now we can define the BERTModel class by instantiating the three classes BERTEncoder, MaskLM, and NextSentencePred. The forward inference returns the encoded BERT representations encoded_X, predictions of masked language modeling mlm_Y_hat, and next sentence predictions nsp_Y_hat.

```
#@save
class BERTModel(nn.Module):
    """The BERT model."""
    def __init__(self, vocab_size, num_hiddens, ffn_num_hiddens,
                 num_heads, num_blks, dropout, max_len=1000):
        super(BERTModel, self).__init__()
        self.encoder = BERTEncoder(vocab_size, num_hiddens, ffn_num_hiddens,
                                   num_heads, num_blks, dropout,
                                   max_len=max_len)
        self.hidden = nn.Sequential(nn.LazyLinear(num_hiddens),
                                    nn.Tanh())
        self.mlm = MaskLM(vocab_size, num_hiddens)
        self.nsp = NextSentencePred()

    def forward(self, tokens, segments, valid_lens=None, pred_positions=None):
        encoded_X = self.encoder(tokens, segments, valid_lens)
        if pred_positions is not None:
            mlm_Y_hat = self.mlm(encoded_X, pred_positions)
        else:
            mlm_Y_hat = None
        # The hidden layer of the MLP classifier for next sentence prediction.
        # 0 is the index of the '<cls>' token
        nsp_Y_hat = self.nsp(self.hidden(encoded_X[:, 0, :]))
        return encoded_X, mlm_Y_hat, nsp_Y_hat
```

## 15.8.7 Summary

- Word embedding models such as word2vec and GloVe are context-independent. They assign the same pretrained vector to the same word regardless of the context of the word (if any). It is hard for them to handle well polysemy or complex semantics in natural languages.
- For context-sensitive word representations such as ELMo and GPT, representations of words depend on their contexts.
- ELMo encodes context bidirectionally but uses task-specific architectures (however, it is practically non-trivial to craft a specific architecture for every natural language processing task); while GPT is task-agnostic but encodes context left-to-right.
- BERT combines the best of both worlds: it encodes context bidirectionally and requires minimal architecture changes for a wide range of natural language processing tasks.
- The embeddings of the BERT input sequence are the sum of the token embeddings, segment embeddings, and positional embeddings.
- Pretraining BERT is composed of two tasks: masked language modeling and next sentence prediction. The former is able to encode bidirectional context for representing words, while the latter explicitly models the logical relationship between text pairs.

### 15.8.8 Exercises

1. All other things being equal, will a masked language model require more or fewer pretraining steps to converge than a left-to-right language model? Why?
2. In the original implementation of BERT, the positionwise feed-forward network in `BERTEncoder` (via `d2l.TransformerEncoderBlock`) and the fully connected layer in `MaskLM` both use the Gaussian error linear unit (GELU) (Hendrycks and Gimpel, 2016) as the activation function. Research into the difference between GELU and ReLU.

Discussions[238].

238

## 15.9 The Dataset for Pretraining BERT

To pretrain the BERT model as implemented in Section 15.8, we need to generate the dataset in the ideal format to facilitate the two pretraining tasks: masked language modeling and next sentence prediction. On the one hand, the original BERT model is pretrained on the concatenation of two huge corpora BookCorpus and English Wikipedia (see Section 15.8.5), making it hard to run for most readers of this book. On the other hand, the off-the-shelf pretrained BERT model may not fit for applications from specific domains like medicine. Thus, it is getting popular to pretrain BERT on a customized dataset. To facilitate the demonstration of BERT pretraining, we use a smaller corpus WikiText-2 (Merity *et al.*, 2016).

Comparing with the PTB dataset used for pretraining word2vec in Section 15.3, WikiText-2 (i) retains the original punctuation, making it suitable for next sentence prediction; (ii) retains the original case and numbers; (iii) is over twice larger.

```
import os
import random
import torch
from d2l import torch as d2l
```

In the WikiText-2 dataset, each line represents a paragraph where space is inserted between any punctuation and its preceding token. Paragraphs with at least two sentences are retained. To split sentences, we only use the period as the delimiter for simplicity. We leave discussions of more complex sentence splitting techniques in the exercises at the end of this section.

```
#@save
d2l.DATA_HUB['wikitext-2'] = (
    'https://s3.amazonaws.com/research.metamind.io/wikitext/'
    'wikitext-2-v1.zip', '3c914d17d80b1459be871a5039ac23e752a53cbe')
```

```
#@save
def _read_wiki(data_dir):
    file_name = os.path.join(data_dir, 'wiki.train.tokens')
    with open(file_name, 'r') as f:
        lines = f.readlines()
    # Uppercase letters are converted to lowercase ones
    paragraphs = [line.strip().lower().split(' . ')
                  for line in lines if len(line.split(' . ')) >= 2]
    random.shuffle(paragraphs)
    return paragraphs
```

## 15.9.1 Defining Helper Functions for Pretraining Tasks

In the following, we begin by implementing helper functions for the two BERT pretraining tasks: next sentence prediction and masked language modeling. These helper functions will be invoked later when transforming the raw text corpus into the dataset of the ideal format to pretrain BERT.

### Generating the Next Sentence Prediction Task

According to descriptions of Section 15.8.5, the `_get_next_sentence` function generates a training example for the binary classification task.

```
#@save
def _get_next_sentence(sentence, next_sentence, paragraphs):
    if random.random() < 0.5:
        is_next = True
    else:
        # `paragraphs` is a list of lists of lists
        next_sentence = random.choice(random.choice(paragraphs))
        is_next = False
    return sentence, next_sentence, is_next
```

The following function generates training examples for next sentence prediction from the input `paragraph` by invoking the `_get_next_sentence` function. Here `paragraph` is a list of sentences, where each sentence is a list of tokens. The argument `max_len` specifies the maximum length of a BERT input sequence during pretraining.

```
#@save
def _get_nsp_data_from_paragraph(paragraph, paragraphs, vocab, max_len):
    nsp_data_from_paragraph = []
    for i in range(len(paragraph) - 1):
        tokens_a, tokens_b, is_next = _get_next_sentence(
            paragraph[i], paragraph[i + 1], paragraphs)
        # Consider 1 '<cls>' token and 2 '<sep>' tokens
        if len(tokens_a) + len(tokens_b) + 3 > max_len:
```

```
            continue
        tokens, segments = d2l.get_tokens_and_segments(tokens_a, tokens_b)
        nsp_data_from_paragraph.append((tokens, segments, is_next))
    return nsp_data_from_paragraph
```

### Generating the Masked Language Modeling Task

In order to generate training examples for the masked language modeling task from a BERT input sequence, we define the following `_replace_mlm_tokens` function. In its inputs, `tokens` is a list of tokens representing a BERT input sequence, `candidate_pred_positions` is a list of token indices of the BERT input sequence excluding those of special tokens (special tokens are not predicted in the masked language modeling task), and `num_mlm_preds` indicates the number of predictions (recall 15% random tokens to predict). Following the definition of the masked language modeling task in Section 15.8.5, at each prediction position, the input may be replaced by a special “<mask>” token or a random token, or remain unchanged. In the end, the function returns the input tokens after possible replacement, the token indices where predictions take place and labels for these predictions.

```
#@save
def _replace_mlm_tokens(tokens, candidate_pred_positions, num_mlm_preds,
                        vocab):
    # For the input of a masked language model, make a new copy of tokens and
    # replace some of them by '<mask>' or random tokens
    mlm_input_tokens = [token for token in tokens]
    pred_positions_and_labels = []
    # Shuffle for getting 15% random tokens for prediction in the masked
    # language modeling task
    random.shuffle(candidate_pred_positions)
    for mlm_pred_position in candidate_pred_positions:
        if len(pred_positions_and_labels) >= num_mlm_preds:
            break
        masked_token = None
        # 80% of the time: replace the word with the '<mask>' token
        if random.random() < 0.8:
            masked_token = '<mask>'
        else:
            # 10% of the time: keep the word unchanged
            if random.random() < 0.5:
                masked_token = tokens[mlm_pred_position]
            # 10% of the time: replace the word with a random word
            else:
                masked_token = random.choice(vocab.idx_to_token)
        mlm_input_tokens[mlm_pred_position] = masked_token
        pred_positions_and_labels.append(
            (mlm_pred_position, tokens[mlm_pred_position]))
    return mlm_input_tokens, pred_positions_and_labels
```

By invoking the aforementioned `_replace_mlm_tokens` function, the following function takes a BERT input sequence (`tokens`) as an input and returns indices of the input tokens (after possible token replacement as described in Section 15.8.5), the token indices where predictions take place, and label indices for these predictions.

```
#@save
def _get_mlm_data_from_tokens(tokens, vocab):
    candidate_pred_positions = []
    # `tokens` is a list of strings
    for i, token in enumerate(tokens):
        # Special tokens are not predicted in the masked language modeling
        # task
        if token in ['<cls>', '<sep>']:
            continue
        candidate_pred_positions.append(i)
    # 15% of random tokens are predicted in the masked language modeling task
    num_mlm_preds = max(1, round(len(tokens) * 0.15))
    mlm_input_tokens, pred_positions_and_labels = _replace_mlm_tokens(
        tokens, candidate_pred_positions, num_mlm_preds, vocab)
    pred_positions_and_labels = sorted(pred_positions_and_labels,
                                       key=lambda x: x[0])
    pred_positions = [v[0] for v in pred_positions_and_labels]
    mlm_pred_labels = [v[1] for v in pred_positions_and_labels]
    return vocab[mlm_input_tokens], pred_positions, vocab[mlm_pred_labels]
```

## 15.9.2 Transforming Text into the Pretraining Dataset

Now we are almost ready to customize a `Dataset` class for pretraining BERT. Before that, we still need to define a helper function `_pad_bert_inputs` to append the special "<pad>" tokens to the inputs. Its argument `examples` contain the outputs from the helper functions `_get_nsp_data_from_paragraph` and `_get_mlm_data_from_tokens` for the two pretraining tasks.

```
#@save
def _pad_bert_inputs(examples, max_len, vocab):
    max_num_mlm_preds = round(max_len * 0.15)
    all_token_ids, all_segments, valid_lens,  = [], [], []
    all_pred_positions, all_mlm_weights, all_mlm_labels = [], [], []
    nsp_labels = []
    for (token_ids, pred_positions, mlm_pred_label_ids, segments,
         is_next) in examples:
        all_token_ids.append(torch.tensor(token_ids + [vocab['<pad>']] * (
            max_len - len(token_ids)), dtype=torch.long))
        all_segments.append(torch.tensor(segments + [0] * (
            max_len - len(segments)), dtype=torch.long))
        # `valid_lens` excludes count of '<pad>' tokens
        valid_lens.append(torch.tensor(len(token_ids), dtype=torch.float32))
        all_pred_positions.append(torch.tensor(pred_positions + [0] * (
            max_num_mlm_preds - len(pred_positions)), dtype=torch.long))
        # Predictions of padded tokens will be filtered out in the loss via
        # multiplication of 0 weights
        all_mlm_weights.append(
            torch.tensor([1.0] * len(mlm_pred_label_ids) + [0.0] * (
                max_num_mlm_preds - len(pred_positions)),
```

```
            dtype=torch.float32))
        all_mlm_labels.append(torch.tensor(mlm_pred_label_ids + [0] * (
            max_num_mlm_preds - len(mlm_pred_label_ids)), dtype=torch.long))
        nsp_labels.append(torch.tensor(is_next, dtype=torch.long))
    return (all_token_ids, all_segments, valid_lens, all_pred_positions,
            all_mlm_weights, all_mlm_labels, nsp_labels)
```

Putting the helper functions for generating training examples of the two pretraining tasks, and the helper function for padding inputs together, we customize the following `_WikiTextDataset` class as the WikiText-2 dataset for pretraining BERT. By implementing the `__getitem__`function, we can arbitrarily access the pretraining (masked language modeling and next sentence prediction) examples generated from a pair of sentences from the WikiText-2 corpus.

The original BERT model uses WordPiece embeddings whose vocabulary size is 30000 (Wu *et al.*, 2016). The tokenization method of WordPiece is a slight modification of the original byte pair encoding algorithm in Section 15.6.2. For simplicity, we use the `d2l.tokenize` function for tokenization. Infrequent tokens that appear less than five times are filtered out.

```
#@save
class _WikiTextDataset(torch.utils.data.Dataset):
    def __init__(self, paragraphs, max_len):
        # Input `paragraphs[i]` is a list of sentence strings representing a
        # paragraph; while output `paragraphs[i]` is a list of sentences
        # representing a paragraph, where each sentence is a list of tokens
        paragraphs = [d2l.tokenize(
            paragraph, token='word') for paragraph in paragraphs]
        sentences = [sentence for paragraph in paragraphs
                     for sentence in paragraph]
        self.vocab = d2l.Vocab(sentences, min_freq=5, reserved_tokens=[
            '<pad>', '<mask>', '<cls>', '<sep>'])
        # Get data for the next sentence prediction task
        examples = []
        for paragraph in paragraphs:
            examples.extend(_get_nsp_data_from_paragraph(
                paragraph, paragraphs, self.vocab, max_len))
        # Get data for the masked language model task
        examples = [(_get_mlm_data_from_tokens(tokens, self.vocab)
                      + (segments, is_next))
                     for tokens, segments, is_next in examples]
        # Pad inputs
        (self.all_token_ids, self.all_segments, self.valid_lens,
         self.all_pred_positions, self.all_mlm_weights,
         self.all_mlm_labels, self.nsp_labels) = _pad_bert_inputs(
            examples, max_len, self.vocab)

    def __getitem__(self, idx):
        return (self.all_token_ids[idx], self.all_segments[idx],
                self.valid_lens[idx], self.all_pred_positions[idx],
                self.all_mlm_weights[idx], self.all_mlm_labels[idx],
```

```
                self.nsp_labels[idx])

    def __len__(self):
        return len(self.all_token_ids)
```

By using the `_read_wiki` function and the `_WikiTextDataset` class, we define the following `load_data_wiki` to download and WikiText-2 dataset and generate pretraining examples from it.

```
#@save
def load_data_wiki(batch_size, max_len):
    """Load the WikiText-2 dataset."""
    num_workers = d2l.get_dataloader_workers()
    data_dir = d2l.download_extract('wikitext-2', 'wikitext-2')
    paragraphs = _read_wiki(data_dir)
    train_set = _WikiTextDataset(paragraphs, max_len)
    train_iter = torch.utils.data.DataLoader(train_set, batch_size,
                                        shuffle=True, num_workers=num_workers)
    return train_iter, train_set.vocab
```

Setting the batch size to 512 and the maximum length of a BERT input sequence to be 64, we print out the shapes of a minibatch of BERT pretraining examples. Note that in each BERT input sequence, $10$ ($64 \times 0.15$) positions are predicted for the masked language modeling task.

```
batch_size, max_len = 512, 64
train_iter, vocab = load_data_wiki(batch_size, max_len)

for (tokens_X, segments_X, valid_lens_x, pred_positions_X, mlm_weights_X,
     mlm_Y, nsp_y) in train_iter:
    print(tokens_X.shape, segments_X.shape, valid_lens_x.shape,
          pred_positions_X.shape, mlm_weights_X.shape, mlm_Y.shape,
          nsp_y.shape)
    break
```

```
torch.Size([512, 64]) torch.Size([512, 64]) torch.Size([512]) torch.Size([512,
↪10]) torch.Size([512, 10]) torch.Size([512, 10]) torch.Size([512])
```

In the end, let's take a look at the vocabulary size. Even after filtering out infrequent tokens, it is still over twice larger than that of the PTB dataset.

```
len(vocab)
```

```
20256
```

### 15.9.3 Summary

- Comparing with the PTB dataset, the WikiText-2 dateset retains the original punctuation, case and numbers, and is over twice larger.
- We can arbitrarily access the pretraining (masked language modeling and next sentence prediction) examples generated from a pair of sentences from the WikiText-2 corpus.

### 15.9.4 Exercises

1. For simplicity, the period is used as the only delimiter for splitting sentences. Try other sentence splitting techniques, such as the spaCy and NLTK. Take NLTK as an example. You need to install NLTK first: `pip install nltk`. In the code, first `import nltk`. Then, download the Punkt sentence tokenizer: `nltk.download('punkt')`. To split sentences such as `sentences = 'This is great ! Why not ?'`, invoking `nltk.tokenize.sent_tokenize(sentences)` will return a list of two sentence strings: `['This is great !', 'Why not ?']`.
2. What is the vocabulary size if we do not filter out any infrequent token?

Discussions[239].

## 15.10 Pretraining BERT

With the BERT model implemented in Section 15.8 and the pretraining examples generated from the WikiText-2 dataset in Section 15.9, we will pretrain BERT on the WikiText-2 dataset in this section.

```
import torch
from torch import nn
from d2l import torch as d2l
```

To start, we load the WikiText-2 dataset as minibatches of pretraining examples for masked language modeling and next sentence prediction. The batch size is 512 and the maximum length of a BERT input sequence is 64. Note that in the original BERT model, the maximum length is 512.

```
batch_size, max_len = 512, 64
train_iter, vocab = d2l.load_data_wiki(batch_size, max_len)
```

### 15.10.1 Pretraining BERT

The original BERT has two versions of different model sizes (Devlin *et al.*, 2018). The base model ($\text{BERT}_{\text{BASE}}$) uses 12 layers (Transformer encoder blocks) with 768 hidden units (hidden size) and 12 self-attention heads. The large model ($\text{BERT}_{\text{LARGE}}$) uses 24 layers with 1024 hidden units and 16 self-attention heads. Notably, the former has 110 million parameters while the latter has 340 million parameters. For demonstration with ease, we define a small BERT, using 2 layers, 128 hidden units, and 2 self-attention heads.

```
net = d2l.BERTModel(len(vocab), num_hiddens=128,
                    ffn_num_hiddens=256, num_heads=2, num_blks=2, dropout=0.2)
devices = d2l.try_all_gpus()
loss = nn.CrossEntropyLoss()
```

Before defining the training loop, we define a helper function `_get_batch_loss_bert`. Given the shard of training examples, this function computes the loss for both the masked language modeling and next sentence prediction tasks. Note that the final loss of BERT pretraining is just the sum of both the masked language modeling loss and the next sentence prediction loss.

```
#@save
def _get_batch_loss_bert(net, loss, vocab_size, tokens_X,
                         segments_X, valid_lens_x,
                         pred_positions_X, mlm_weights_X,
                         mlm_Y, nsp_y):
    # Forward pass
    _, mlm_Y_hat, nsp_Y_hat = net(tokens_X, segments_X,
                                  valid_lens_x.reshape(-1),
                                  pred_positions_X)
    # Compute masked language model loss
    mlm_l = loss(mlm_Y_hat.reshape(-1, vocab_size), mlm_Y.reshape(-1)) *\
    mlm_weights_X.reshape(-1, 1)
    mlm_l = mlm_l.sum() / (mlm_weights_X.sum() + 1e-8)
    # Compute next sentence prediction loss
    nsp_l = loss(nsp_Y_hat, nsp_y)
    l = mlm_l + nsp_l
    return mlm_l, nsp_l, l
```

Invoking the two aforementioned helper functions, the following `train_bert` function defines the procedure to pretrain BERT (`net`) on the WikiText-2 (`train_iter`) dataset. Training BERT can take very long. Instead of specifying the number of epochs for training as in the `train_ch13` function (see Section 14.1), the input `num_steps` of the following function specifies the number of iteration steps for training.

```
def train_bert(train_iter, net, loss, vocab_size, devices, num_steps):
    net(*next(iter(train_iter))[:4])
    net = nn.DataParallel(net, device_ids=devices).to(devices[0])
    trainer = torch.optim.Adam(net.parameters(), lr=0.01)
    step, timer = 0, d2l.Timer()
    animator = d2l.Animator(xlabel='step', ylabel='loss',
                            xlim=[1, num_steps], legend=['mlm', 'nsp'])
```

```
    # Sum of masked language modeling losses, sum of next sentence prediction
    # losses, no. of sentence pairs, count
    metric = d2l.Accumulator(4)
    num_steps_reached = False
    while step < num_steps and not num_steps_reached:
        for tokens_X, segments_X, valid_lens_x, pred_positions_X,\
            mlm_weights_X, mlm_Y, nsp_y in train_iter:
            tokens_X = tokens_X.to(devices[0])
            segments_X = segments_X.to(devices[0])
            valid_lens_x = valid_lens_x.to(devices[0])
            pred_positions_X = pred_positions_X.to(devices[0])
            mlm_weights_X = mlm_weights_X.to(devices[0])
            mlm_Y, nsp_y = mlm_Y.to(devices[0]), nsp_y.to(devices[0])
            trainer.zero_grad()
            timer.start()
            mlm_l, nsp_l, l = _get_batch_loss_bert(
                net, loss, vocab_size, tokens_X, segments_X, valid_lens_x,
                pred_positions_X, mlm_weights_X, mlm_Y, nsp_y)
            l.backward()
            trainer.step()
            metric.add(mlm_l, nsp_l, tokens_X.shape[0], 1)
            timer.stop()
            animator.add(step + 1,
                         (metric[0] / metric[3], metric[1] / metric[3]))
            step += 1
            if step == num_steps:
                num_steps_reached = True
                break

    print(f'MLM loss {metric[0] / metric[3]:.3f}, '
          f'NSP loss {metric[1] / metric[3]:.3f}')
    print(f'{metric[2] / timer.sum():.1f} sentence pairs/sec on '
          f'{str(devices)}')
```

We can plot both the masked language modeling loss and the next sentence prediction loss during BERT pretraining.

```
train_bert(train_iter, net, loss, len(vocab), devices, 50)
```

```
MLM loss 5.411, NSP loss 0.764
2447.5 sentence pairs/sec on [device(type='cuda', index=0), device(type='cuda',
↪ index=1)]
```

### 15.10.2 Representing Text with BERT

After pretraining BERT, we can use it to represent single text, text pairs, or any token in them. The following function returns the BERT (`net`) representations for all tokens in `tokens_a` and `tokens_b`.

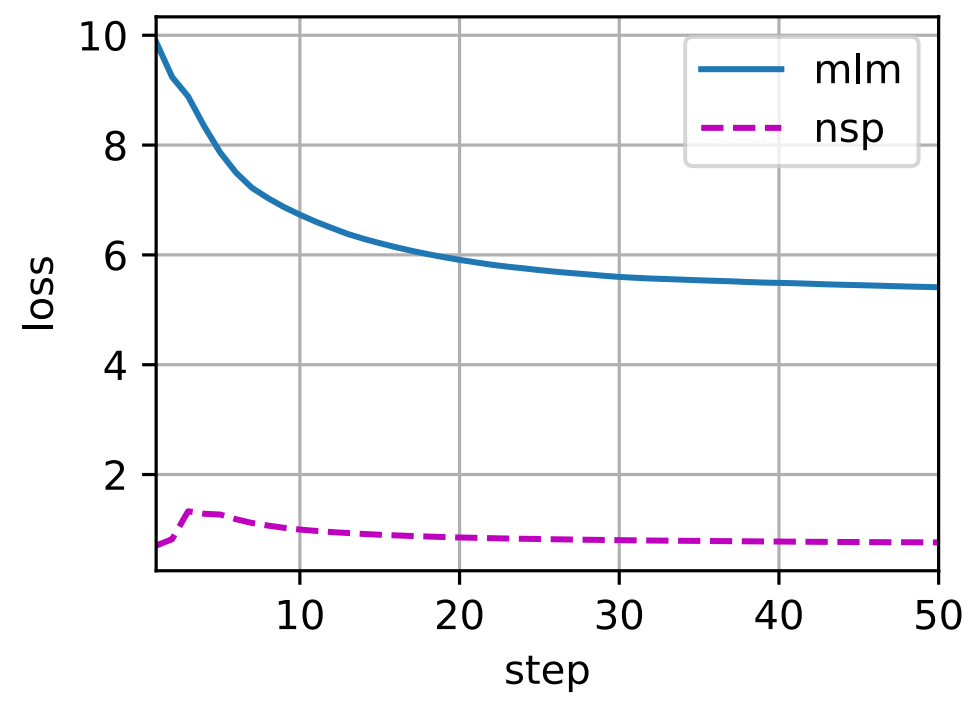

```
def get_bert_encoding(net, tokens_a, tokens_b=None):
    tokens, segments = d2l.get_tokens_and_segments(tokens_a, tokens_b)
    token_ids = torch.tensor(vocab[tokens], device=devices[0]).unsqueeze(0)
    segments = torch.tensor(segments, device=devices[0]).unsqueeze(0)
    valid_len = torch.tensor(len(tokens), device=devices[0]).unsqueeze(0)
    encoded_X, _, _ = net(token_ids, segments, valid_len)
    return encoded_X
```

Consider the sentence "a crane is flying". Recall the input representation of BERT as discussed in Section 15.8.4. After inserting special tokens "<cls>" (used for classification) and "<sep>" (used for separation), the BERT input sequence has a length of six. Since zero is the index of the "<cls>" token, `encoded_text[:, 0, :]` is the BERT representation of the entire input sentence. To evaluate the polysemy token "crane", we also print out the first three elements of the BERT representation of the token.

```
tokens_a = ['a', 'crane', 'is', 'flying']
encoded_text = get_bert_encoding(net, tokens_a)
# Tokens: '<cls>', 'a', 'crane', 'is', 'flying', '<sep>'
encoded_text_cls = encoded_text[:, 0, :]
encoded_text_crane = encoded_text[:, 2, :]
encoded_text.shape, encoded_text_cls.shape, encoded_text_crane[0][:3]
```

```
(torch.Size([1, 6, 128]),
 torch.Size([1, 128]),
 tensor([ 0.0282, -0.6125, -0.3779], device='cuda:0', grad_fn=<SliceBackward0>
↪))
```

Now consider a sentence pair "a crane driver came" and "he just left". Similarly, `encoded_pair[:, 0, :]` is the encoded result of the entire sentence pair from the pretrained BERT. Note that the first three elements of the polysemy token "crane" are different from those when the context is different. This supports that BERT representations are context-sensitive.

```
tokens_a, tokens_b = ['a', 'crane', 'driver', 'came'], ['he', 'just', 'left']
encoded_pair = get_bert_encoding(net, tokens_a, tokens_b)
```

```
# Tokens: '<cls>', 'a', 'crane', 'driver', 'came', '<sep>', 'he', 'just',
# 'left', '<sep>'
encoded_pair_cls = encoded_pair[:, 0, :]
encoded_pair_crane = encoded_pair[:, 2, :]
encoded_pair.shape, encoded_pair_cls.shape, encoded_pair_crane[0][:3]
```

```
(torch.Size([1, 10, 128]),
 torch.Size([1, 128]),
 tensor([-1.3067, -0.3663,  0.0759], device='cuda:0', grad_fn=<SliceBackward0>
↪))
```

In Chapter 16, we will fine-tune a pretrained BERT model for downstream natural language processing applications.

### 15.10.3 Summary

- The original BERT has two versions, where the base model has 110 million parameters and the large model has 340 million parameters.
- After pretraining BERT, we can use it to represent single text, text pairs, or any token in them.
- In the experiment, the same token has different BERT representation when their contexts are different. This supports that BERT representations are context-sensitive.

### 15.10.4 Exercises

1. In the experiment, we can see that the masked language modeling loss is significantly higher than the next sentence prediction loss. Why?
2. Set the maximum length of a BERT input sequence to be 512 (same as the original BERT model). Use the configurations of the original BERT model such as $\text{BERT}_{\text{LARGE}}$. Do you encounter any error when running this section? Why?

Discussions[240].

# 16 Natural Language Processing: Applications

We have seen how to represent tokens in text sequences and train their representations in Chapter 15. Such pretrained text representations can be fed to various models for different downstream natural language processing tasks.

In fact, earlier chapters have already discussed some natural language processing applications *without pretraining*, just for explaining deep learning architectures. For instance, in Chapter 9, we have relied on RNNs to design language models to generate novella-like text. In Chapter 10 and Chapter 11, we have also designed models based on RNNs and attention mechanisms for machine translation.

However, this book does not intend to cover all such applications in a comprehensive manner. Instead, our focus is on *how to apply (deep) representation learning of languages to addressing natural language processing problems*. Given pretrained text representations, this chapter will explore two popular and representative downstream natural language processing tasks: sentiment analysis and natural language inference, which analyze single text and relationships of text pairs, respectively.

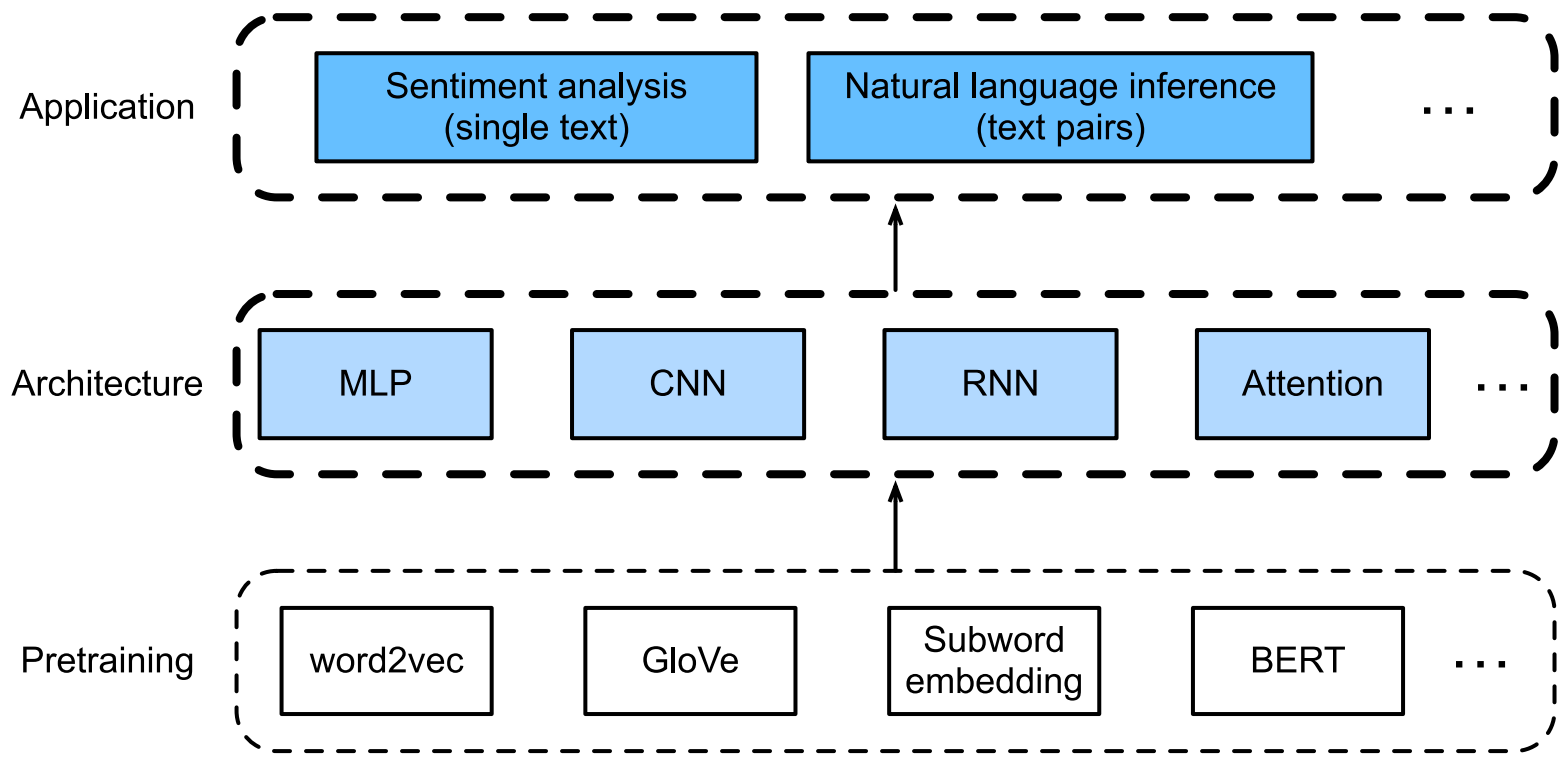

Fig. 16.1 Pretrained text representations can be fed to various deep learning architectures for different downstream natural language processing applications. This chapter focuses on how to design models for different downstream natural language processing applications.

As depicted in Fig. 16.1, this chapter focuses on describing the basic ideas of designing natural language processing models using different types of deep learning architectures, such as MLPs, CNNs, RNNs, and attention. Though it is possible to combine any pretrained text representations with any architecture for either application in Fig. 16.1, we select a few rep-resentative combinations. Specifically, we will explore popular architectures based on RNNs and CNNs for sentiment analysis. For natural language inference, we choose attention and MLPs to demonstrate how to analyze text pairs. In the end, we introduce how to fine-tune a pretrained BERT model for a wide range of natural language processing applications, such as on a sequence level (single text classification and text pair classification) and a token level (text tagging and question answering). As a concrete empirical case, we will fine-tune BERT for natural language inference.

As we have introduced in Section 15.8, BERT requires minimal architecture changes for a wide range of natural language processing applications. However, this benefit comes at the cost of fine-tuning a huge number of BERT parameters for the downstream applications. When space or time is limited, those crafted models based on MLPs, CNNs, RNNs, and attention are more feasible. In the following, we start by the sentiment analysis application and illustrate the model design based on RNNs and CNNs, respectively.

## 16.1 Sentiment Analysis and the Dataset

With the proliferation of online social media and review platforms, a plethora of opinionated data has been logged, bearing great potential for supporting decision making processes. *Sentiment analysis* studies people's sentiments in their produced text, such as product reviews, blog comments, and forum discussions. It enjoys wide applications to fields as diverse as politics (e.g., analysis of public sentiments towards policies), finance (e.g., analysis of sentiments of the market), and marketing (e.g., product research and brand management).

Since sentiments can be categorized as discrete polarities or scales (e.g., positive and negative), we can consider sentiment analysis as a text classification task, which transforms a varying-length text sequence into a fixed-length text category. In this chapter, we will use Stanford's large movie review dataset[241] for sentiment analysis. It consists of a training set and a testing set, either containing 25000 movie reviews downloaded from IMDb. In both datasets, there are equal number of "positive" and "negative" labels, indicating different sentiment polarities.

```
import os
import torch
from torch import nn
from d2l import torch as d2l
```

### 16.1.1 Reading the Dataset

First, download and extract this IMDb review dataset in the path `../data/aclImdb`.

```
#@save
d2l.DATA_HUB['aclImdb'] = (d2l.DATA_URL + 'aclImdb_v1.tar.gz',
                          '01ada507287d82875905620988597833ad4e0903')

data_dir = d2l.download_extract('aclImdb', 'aclImdb')
```

Next, read the training and test datasets. Each example is a review and its label: 1 for "positive" and 0 for "negative".

```
#@save
def read_imdb(data_dir, is_train):
    """Read the IMDb review dataset text sequences and labels."""
    data, labels = [], []
    for label in ('pos', 'neg'):
        folder_name = os.path.join(data_dir, 'train' if is_train else 'test',
                                   label)
        for file in os.listdir(folder_name):
            with open(os.path.join(folder_name, file), 'rb') as f:
                review = f.read().decode('utf-8').replace('\n', '')
                data.append(review)
                labels.append(1 if label == 'pos' else 0)
    return data, labels

train_data = read_imdb(data_dir, is_train=True)
print('# trainings:', len(train_data[0]))
for x, y in zip(train_data[0][:3], train_data[1][:3]):
    print('label:', y, 'review:', x[:60])
```

```
# trainings: 25000
label: 1 review: Shocking, well-made chiller is an undervalued tale of atroci
label: 1 review: Fast-paced, funny, sexy, and spectacular. Cagney is always t
label: 1 review: Especially for a time when not much science fiction was bein
```

## 16.1.2 Preprocessing the Dataset

Treating each word as a token and filtering out words that appear less than 5 times, we create a vocabulary out of the training dataset.

```
train_tokens = d2l.tokenize(train_data[0], token='word')
vocab = d2l.Vocab(train_tokens, min_freq=5, reserved_tokens=['<pad>'])
```

After tokenization, let's plot the histogram of review lengths in tokens.

```
d2l.set_figsize()
d2l.plt.xlabel('# tokens per review')
d2l.plt.ylabel('count')
d2l.plt.hist([len(line) for line in train_tokens], bins=range(0, 1000, 50));
```

As we expected, the reviews have varying lengths. To process a minibatch of such reviews at each time, we set the length of each review to 500 with truncation and padding, which is similar to the preprocessing step for the machine translation dataset in Section 10.5.

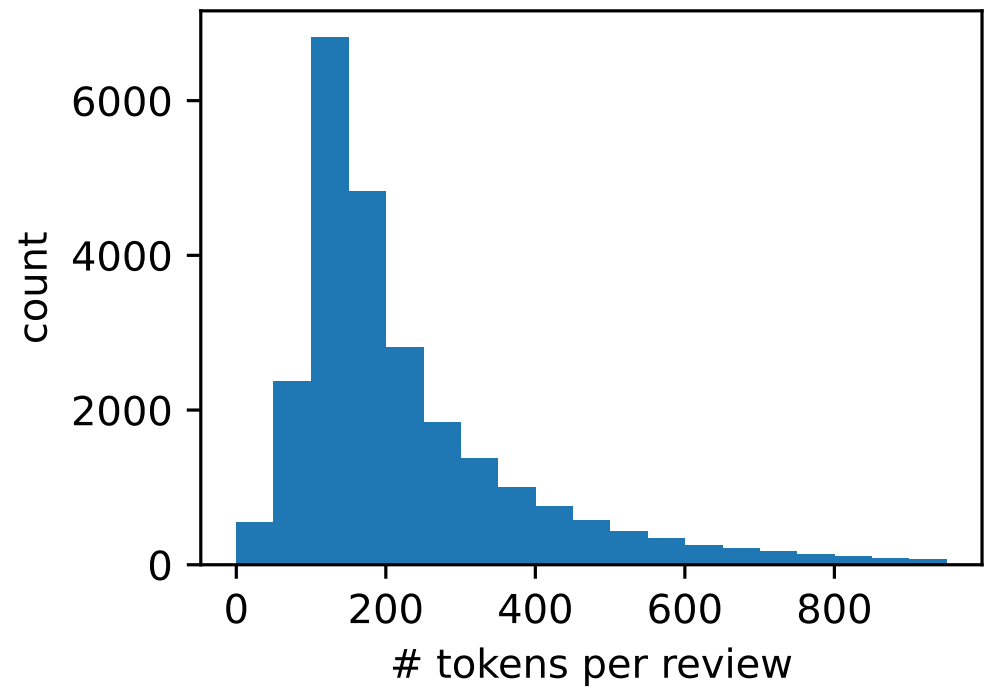

```
num_steps = 500  # sequence length
train_features = torch.tensor([d2l.truncate_pad(
    vocab[line], num_steps, vocab['<pad>']) for line in train_tokens])
print(train_features.shape)
```

```
torch.Size([25000, 500])
```

### 16.1.3 Creating Data Iterators

Now we can create data iterators. At each iteration, a minibatch of examples are returned.

```
train_iter = d2l.load_array((train_features, torch.tensor(train_data[1])), 64)

for X, y in train_iter:
    print('X:', X.shape, ', y:', y.shape)
    break
print('# batches:', len(train_iter))
```

```
X: torch.Size([64, 500]) , y: torch.Size([64])
# batches: 391
```

### 16.1.4 Putting It All Together

Last, we wrap up the above steps into the `load_data_imdb` function. It returns training and test data iterators and the vocabulary of the IMDb review dataset.

```
#@save
def load_data_imdb(batch_size, num_steps=500):
```

```
    """Return data iterators and the vocabulary of the IMDb review dataset."""
    data_dir = d2l.download_extract('aclImdb', 'aclImdb')
    train_data = read_imdb(data_dir, True)
    test_data = read_imdb(data_dir, False)
    train_tokens = d2l.tokenize(train_data[0], token='word')
    test_tokens = d2l.tokenize(test_data[0], token='word')
    vocab = d2l.Vocab(train_tokens, min_freq=5)
    train_features = torch.tensor([d2l.truncate_pad(
        vocab[line], num_steps, vocab['<pad>']) for line in train_tokens])
    test_features = torch.tensor([d2l.truncate_pad(
        vocab[line], num_steps, vocab['<pad>']) for line in test_tokens])
    train_iter = d2l.load_array((train_features, torch.tensor(train_data[1])),
                                batch_size)
    test_iter = d2l.load_array((test_features, torch.tensor(test_data[1])),
                               batch_size,
                               is_train=False)
    return train_iter, test_iter, vocab
```

### 16.1.5 Summary

- Sentiment analysis studies people's sentiments in their produced text, which is considered as a text classification problem that transforms a varying-length text sequence into a fixed-length text category.
- After preprocessing, we can load Stanford's large movie review dataset (IMDb review dataset) into data iterators with a vocabulary.

### 16.1.6 Exercises

1. What hyperparameters in this section can we modify to accelerate training sentiment analysis models?
2. Can you implement a function to load the dataset of Amazon reviews[242] into data iterators and labels for sentiment analysis?

Discussions[243].

## 16.2 Sentiment Analysis: Using Recurrent Neural Networks

Like word similarity and analogy tasks, we can also apply pretrained word vectors to sentiment analysis. Since the IMDb review dataset in Section 16.1 is not very big, using text representations that were pretrained on large-scale corpora may reduce overfitting of the model.

As a specific example illustrated in Fig. 16.2.1, we will represent each token using the pretrained GloVe model, and feed these token representations into a multilayer bidirectional RNN to obtain the text sequence representation, which will be transformed into sentiment analysis outputs (Maas *et al.*, 2011). For the same downstream application, we will consider a different architectural choice later.

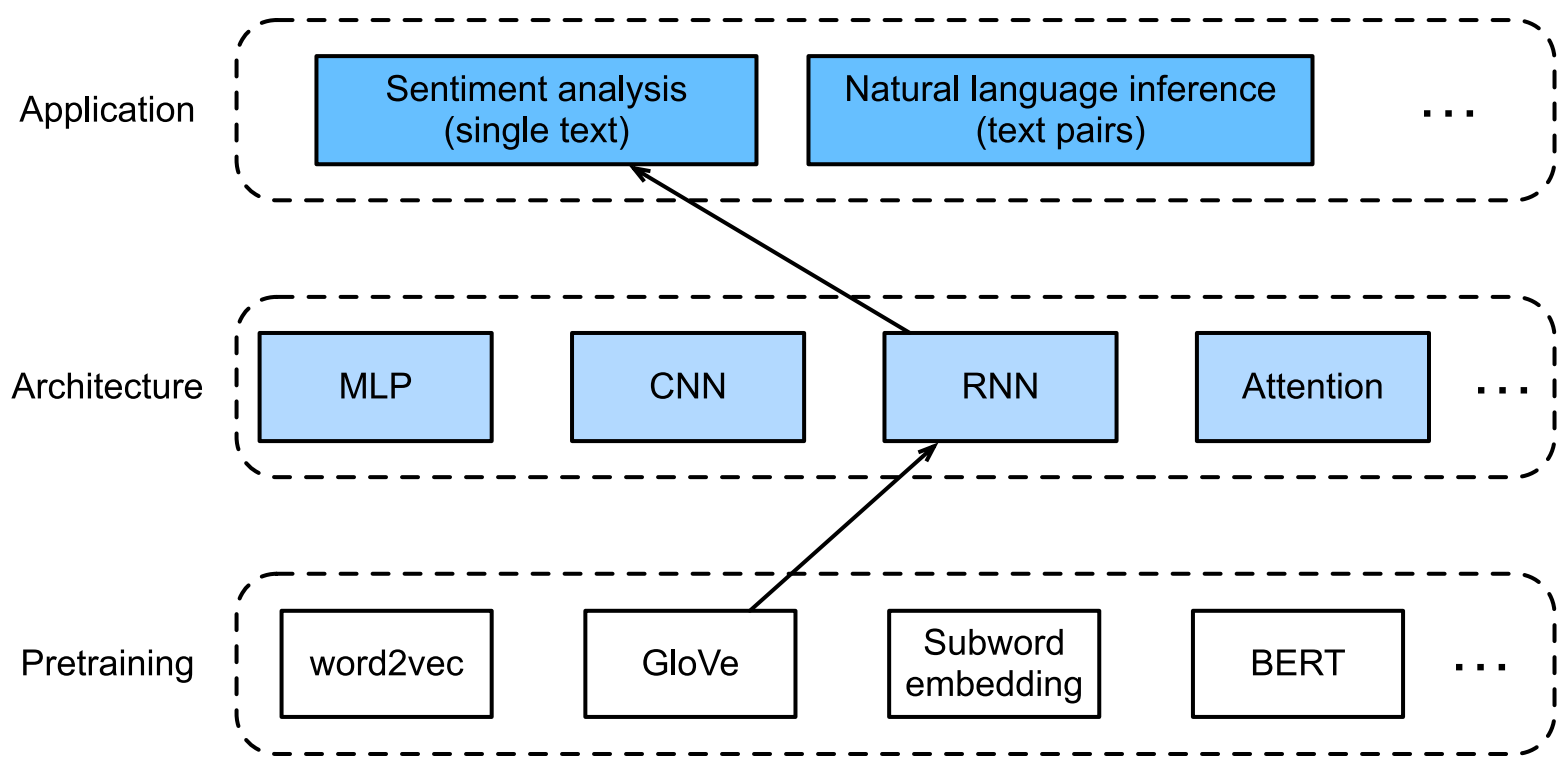

Fig. 16.2.1 This section feeds pretrained GloVe to an RNN-based architecture for sentiment analysis.

```
import torch
from torch import nn
from d2l import torch as d2l

batch_size = 64
train_iter, test_iter, vocab = d2l.load_data_imdb(batch_size)
```

```
Downloading ../data/aclImdb_v1.tar.gz from http://d2l-data.s3-accelerate.
↪amazonaws.com/aclImdb_v1.tar.gz...
```

## 16.2.1 Representing Single Text with RNNs

In text classifications tasks, such as sentiment analysis, a varying-length text sequence will be transformed into fixed-length categories. In the following `BiRNN` class, while each token of a text sequence gets its individual pretrained GloVe representation via the embedding layer (`self.embedding`), the entire sequence is encoded by a bidirectional RNN (`self.encoder`). More concretely, the hidden states (at the last layer) of the bidirectional LSTM at both the initial and final time steps are concatenated as the representation of the text sequence. This single text representation is then transformed into output categories by a fully connected layer (`self.decoder`) with two outputs ("positive" and "negative").

```
class BiRNN(nn.Module):
    def __init__(self, vocab_size, embed_size, num_hiddens,
                 num_layers, **kwargs):
        super(BiRNN, self).__init__(**kwargs)
```

```
        self.embedding = nn.Embedding(vocab_size, embed_size)
        # Set `bidirectional` to True to get a bidirectional RNN
        self.encoder = nn.LSTM(embed_size, num_hiddens, num_layers=num_layers,
                                bidirectional=True)
        self.decoder = nn.Linear(4 * num_hiddens, 2)

    def forward(self, inputs):
        # The shape of `inputs` is (batch size, no. of time steps). Because
        # LSTM requires its input's first dimension to be the temporal
        # dimension, the input is transposed before obtaining token
        # representations. The output shape is (no. of time steps, batch size,
        # word vector dimension)
        embeddings = self.embedding(inputs.T)
        self.encoder.flatten_parameters()
        # Returns hidden states of the last hidden layer at different time
        # steps. The shape of `outputs` is (no. of time steps, batch size,
        # 2 * no. of hidden units)
        outputs, _ = self.encoder(embeddings)
        # Concatenate the hidden states at the initial and final time steps as
        # the input of the fully connected layer. Its shape is (batch size,
        # 4 * no. of hidden units)
        encoding = torch.cat((outputs[0], outputs[-1]), dim=1)
        outs = self.decoder(encoding)
        return outs
```

Let's construct a bidirectional RNN with two hidden layers to represent single text for sentiment analysis.

```
embed_size, num_hiddens, num_layers, devices = 100, 100, 2, d2l.try_all_gpus()
net = BiRNN(len(vocab), embed_size, num_hiddens, num_layers)
```

```
def init_weights(module):
    if type(module) == nn.Linear:
        nn.init.xavier_uniform_(module.weight)
    if type(module) == nn.LSTM:
        for param in module._flat_weights_names:
            if "weight" in param:
                nn.init.xavier_uniform_(module._parameters[param])
net.apply(init_weights);
```

### 16.2.2 Loading Pretrained Word Vectors

Below we load the pretrained 100-dimensional (needs to be consistent with `embed_size`) GloVe embeddings for tokens in the vocabulary.

```
glove_embedding = d2l.TokenEmbedding('glove.6b.100d')
```

```
Downloading ../data/glove.6B.100d.zip from http://d2l-data.s3-accelerate.
↪amazonaws.com/glove.6B.100d.zip...
```

Print the shape of the vectors for all the tokens in the vocabulary.

```
embeds = glove_embedding[vocab.idx_to_token]
embeds.shape
```

```
torch.Size([49346, 100])
```

We use these pretrained word vectors to represent tokens in the reviews and will not update these vectors during training.

```
net.embedding.weight.data.copy_(embeds)
net.embedding.weight.requires_grad = False
```

### 16.2.3 Training and Evaluating the Model

Now we can train the bidirectional RNN for sentiment analysis.

```
lr, num_epochs = 0.01, 5
trainer = torch.optim.Adam(net.parameters(), lr=lr)
loss = nn.CrossEntropyLoss(reduction="none")
d2l.train_ch13(net, train_iter, test_iter, loss, trainer, num_epochs, devices)
```

```
loss 0.693, train acc 0.502, test acc 0.500
2089.4 examples/sec on [device(type='cuda', index=0), device(type='cuda',␣
↪index=1)]
```

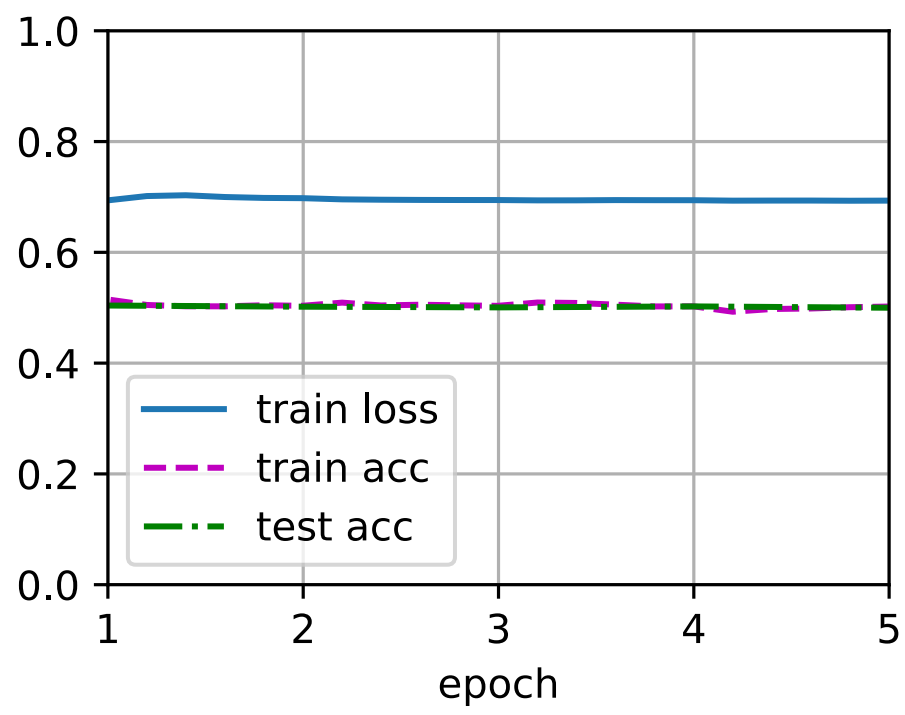

We define the following function to predict the sentiment of a text sequence using the trained model `net`.

```
#@save
def predict_sentiment(net, vocab, sequence):
    """Predict the sentiment of a text sequence."""
    sequence = torch.tensor(vocab[sequence.split()], device=d2l.try_gpu())
    label = torch.argmax(net(sequence.reshape(1, -1)), dim=1)
    return 'positive' if label == 1 else 'negative'
```

Finally, let's use the trained model to predict the sentiment for two simple sentences.

```
predict_sentiment(net, vocab, 'this movie is so great')
```

```
'positive'
```

```
predict_sentiment(net, vocab, 'this movie is so bad')
```

```
'positive'
```

### 16.2.4 Summary

- Pretrained word vectors can represent individual tokens in a text sequence.
- Bidirectional RNNs can represent a text sequence, such as via the concatenation of its hidden states at the initial and final time steps. This single text representation can be transformed into categories using a fully connected layer.

### 16.2.5 Exercises

1. Increase the number of epochs. Can you improve the training and testing accuracies? How about tuning other hyperparameters?
2. Use larger pretrained word vectors, such as 300-dimensional GloVe embeddings. Does it improve classification accuracy?
3. Can we improve the classification accuracy by using the spaCy tokenization? You need to install spaCy (`pip install spacy`) and install the English package (`python -m spacy download en`). In the code, first, import spaCy (`import spacy`). Then, load the spaCy English package (`spacy_en = spacy.load('en')`). Finally, define the function `def tokenizer(text): return [tok.text for tok in spacy_en.tokenizer(text)]` and replace the original `tokenizer` function. Note the different forms of phrase tokens in GloVe and spaCy. For example, the phrase token "new york" takes the form of "new-york" in GloVe and the form of "new york" after the spaCy tokenization.

Discussions[244].

# 16.3 Sentiment Analysis: Using Convolutional Neural Networks

In Chapter 7, we investigated mechanisms for processing two-dimensional image data with two-dimensional CNNs, which were applied to local features such as adjacent pixels. Though originally designed for computer vision, CNNs are also widely used for natural language processing. Simply put, just think of any text sequence as a one-dimensional image. In this way, one-dimensional CNNs can process local features such as $n$-grams in text.

In this section, we will use the *textCNN* model to demonstrate how to design a CNN architecture for representing single text (Kim, 2014). Compared with Fig. 16.2.1 that uses an RNN architecture with GloVe pretraining for sentiment analysis, the only difference in Fig. 16.3.1 lies in the choice of the architecture.

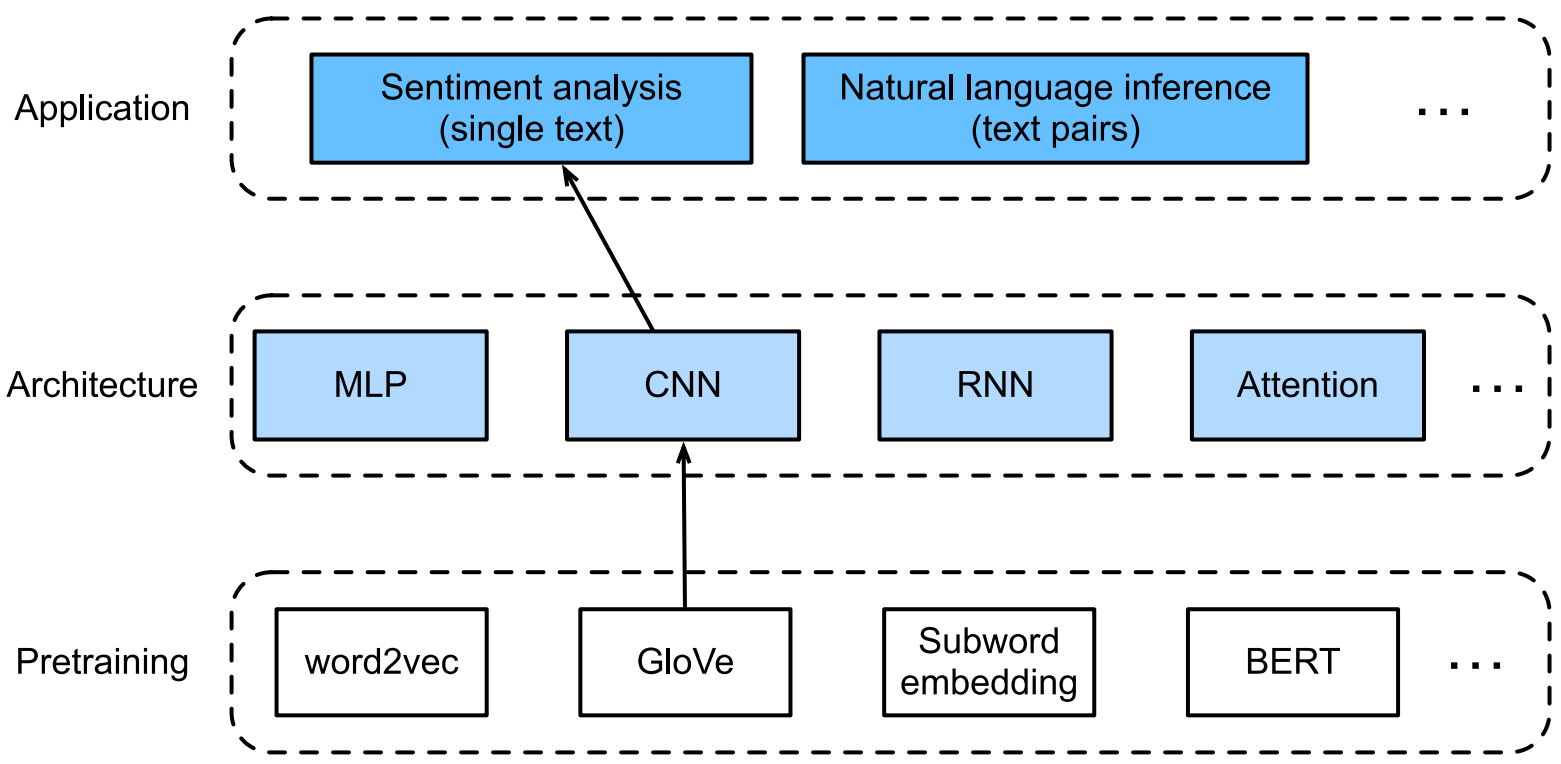

Fig. 16.3.1 This section feeds pretrained GloVe to a CNN-based architecture for sentiment analysis.

```
import torch
from torch import nn
from d2l import torch as d2l

batch_size = 64
train_iter, test_iter, vocab = d2l.load_data_imdb(batch_size)
```

## 16.3.1 One-Dimensional Convolutions

Before introducing the model, let's see how a one-dimensional convolution works. Bear in mind that it is just a special case of a two-dimensional convolution based on the cross-correlation operation.

As shown in Fig. 16.3.2, in the one-dimensional case, the convolution window slides from left to right across the input tensor. During sliding, the input subtensor (e.g., 0 and 1 in Fig.

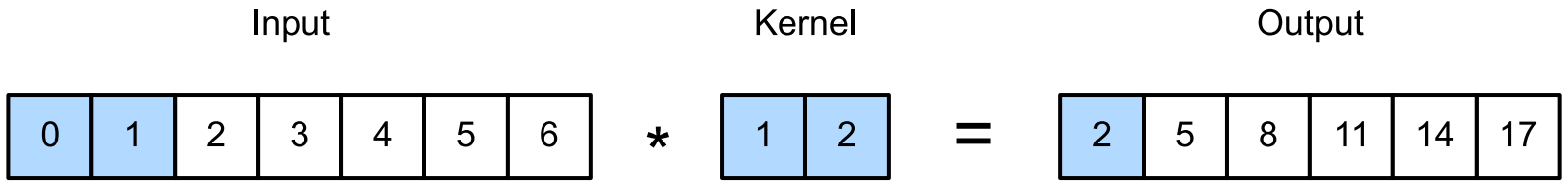

Fig. 16.3.2 One-dimensional cross-correlation operation. The shaded portions are the first output element as well as the input and kernel tensor elements used for the output computation: $0 \times 1 + 1 \times 2 = 2$.

16.3.2) contained in the convolution window at a certain position and the kernel tensor (e.g., 1 and 2 in Fig. 16.3.2) are multiplied elementwise. The sum of these multiplications gives the single scalar value (e.g., $0 \times 1 + 1 \times 2 = 2$ in Fig. 16.3.2) at the corresponding position of the output tensor.

We implement one-dimensional cross-correlation in the following `corr1d` function. Given an input tensor `X` and a kernel tensor `K`, it returns the output tensor `Y`.

```
def corr1d(X, K):
    w = K.shape[0]
    Y = torch.zeros((X.shape[0] - w + 1))
    for i in range(Y.shape[0]):
        Y[i] = (X[i: i + w] * K).sum()
    return Y
```

We can construct the input tensor `X` and the kernel tensor `K` from Fig. 16.3.2 to validate the output of the above one-dimensional cross-correlation implementation.

```
X, K = torch.tensor([0, 1, 2, 3, 4, 5, 6]), torch.tensor([1, 2])
corr1d(X, K)
```

```
tensor([ 2.,  5.,  8., 11., 14., 17.])
```

For any one-dimensional input with multiple channels, the convolution kernel needs to have the same number of input channels. Then for each channel, perform a cross-correlation operation on the one-dimensional tensor of the input and the one-dimensional tensor of the convolution kernel, summing the results over all the channels to produce the one-dimensional output tensor. Fig. 16.3.3 shows a one-dimensional cross-correlation operation with 3 input channels.

We can implement the one-dimensional cross-correlation operation for multiple input channels and validate the results in Fig. 16.3.3.

```
def corr1d_multi_in(X, K):
    # First, iterate through the 0th dimension (channel dimension) of `X` and
    # `K`. Then, add them together
    return sum(corr1d(x, k) for x, k in zip(X, K))

X = torch.tensor([[0, 1, 2, 3, 4, 5, 6],
```

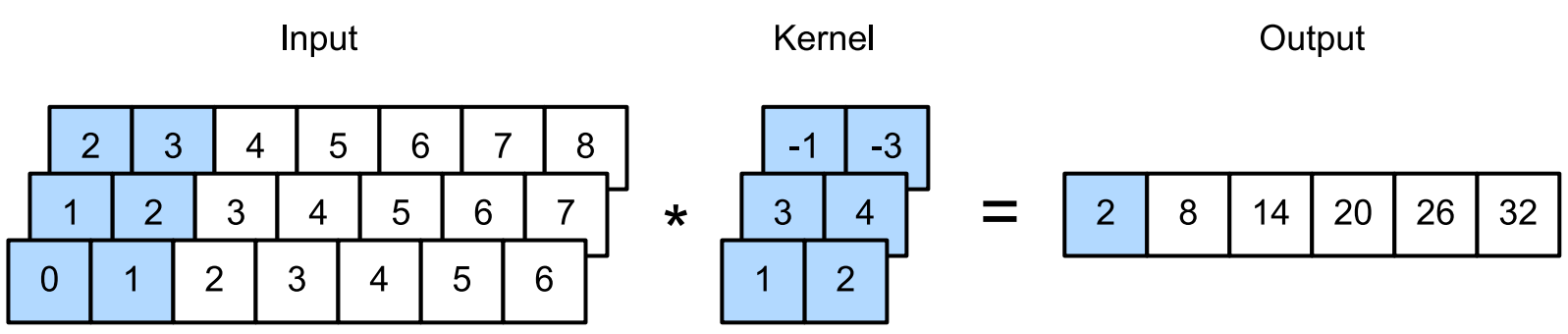

Fig. 16.3.3 One-dimensional cross-correlation operation with 3 input channels. The shaded portions are the first output element as well as the input and kernel tensor elements used for the output computation: $0 \times 1 + 1 \times 2 + 1 \times 3 + 2 \times 4 + 2 \times (-1) + 3 \times (-3) = 2$.

```
              [1, 2, 3, 4, 5, 6, 7],
              [2, 3, 4, 5, 6, 7, 8]])
K = torch.tensor([[1, 2], [3, 4], [-1, -3]])
corr1d_multi_in(X, K)
```

```
tensor([ 2.,  8., 14., 20., 26., 32.])
```

Note that multi-input-channel one-dimensional cross-correlations are equivalent to single-input-channel two-dimensional cross-correlations. To illustrate, an equivalent form of the multi-input-channel one-dimensional cross-correlation in Fig. 16.3.3 is the single-input-channel two-dimensional cross-correlation in Fig. 16.3.4, where the height of the convolution kernel has to be the same as that of the input tensor.

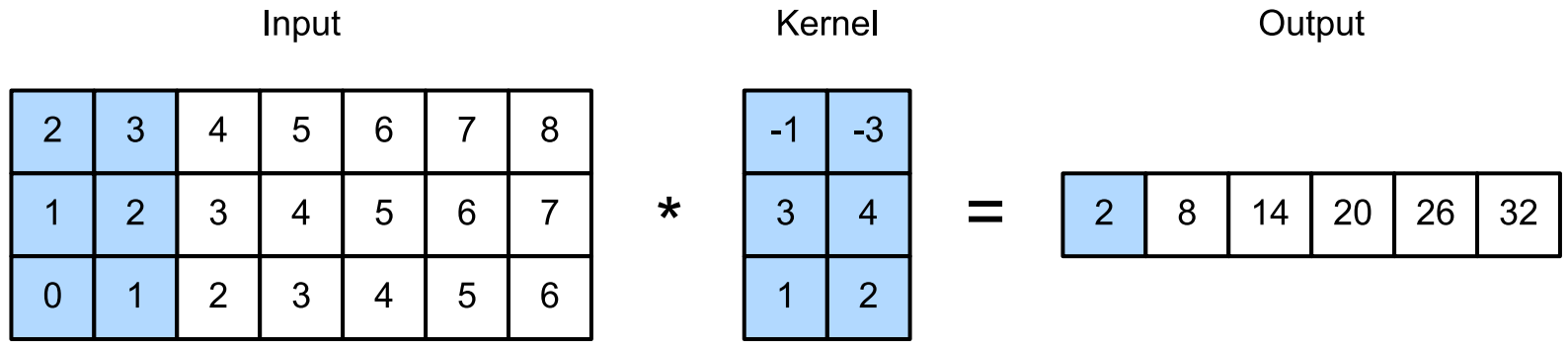

Fig. 16.3.4 Two-dimensional cross-correlation operation with a single input channel. The shaded portions are the first output element as well as the input and kernel tensor elements used for the output computation: $2 \times (-1) + 3 \times (-3) + 1 \times 3 + 2 \times 4 + 0 \times 1 + 1 \times 2 = 2$.

Both the outputs in Fig. 16.3.2 and Fig. 16.3.3 have only one channel. Same as two-dimensional convolutions with multiple output channels described in Section 7.4.2, we can also specify multiple output channels for one-dimensional convolutions.

## 16.3.2 Max-Over-Time Pooling

Similarly, we can use pooling to extract the highest value from sequence representations as the most important feature across time steps. The *max-over-time pooling* used in textCNN works like the one-dimensional global max-pooling (Collobert *et al.*, 2011). For a multi-channel input where each channel stores values at different time steps, the output at each channel is the maximum value for that channel. Note that the max-over-time pooling allows different numbers of time steps at different channels.

## 16.3.3 The textCNN Model

Using the one-dimensional convolution and max-over-time pooling, the textCNN model takes individual pretrained token representations as input, then obtains and transforms sequence representations for the downstream application.

For a single text sequence with $n$ tokens represented by $d$-dimensional vectors, the width, height, and number of channels of the input tensor are $n$, 1, and $d$, respectively. The textCNN model transforms the input into the output as follows:

1. Define multiple one-dimensional convolution kernels and perform convolution operations separately on the inputs. Convolution kernels with different widths may capture local features among different numbers of adjacent tokens.

2. Perform max-over-time pooling on all the output channels, and then concatenate all the scalar pooling outputs as a vector.

3. Transform the concatenated vector into the output categories using the fully connected layer. Dropout can be used for reducing overfitting.

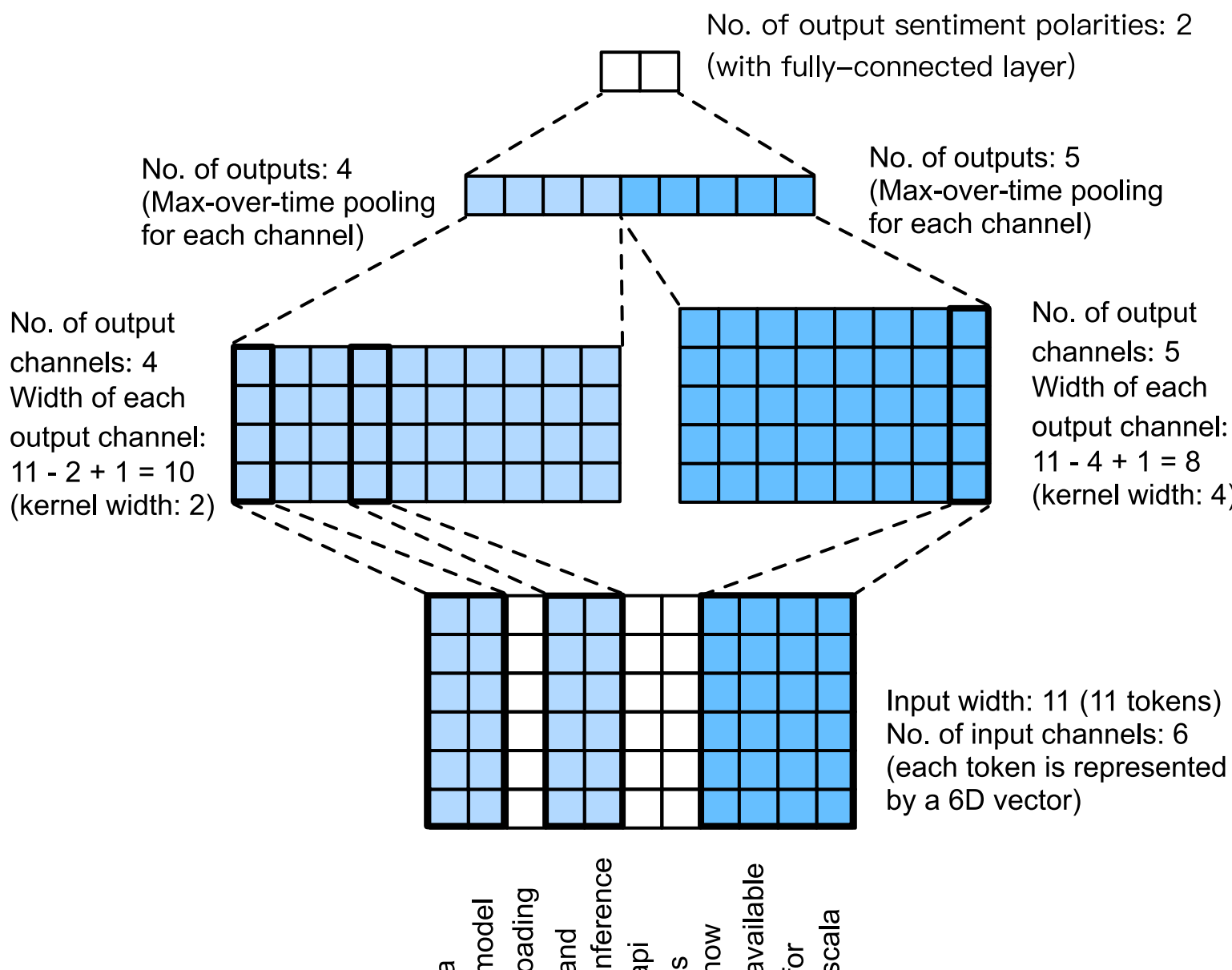

Fig. 16.3.5 The model architecture of textCNN.

Fig. 16.3.5 illustrates the model architecture of textCNN with a concrete example. The input is a sentence with 11 tokens, where each token is represented by a 6-dimensional vectors. So we have a 6-channel input with width 11. Define two one-dimensional convolution kernels of widths 2 and 4, with 4 and 5 output channels, respectively. They produce 4 output channels with width $11-2+1 = 10$ and 5 output channels with width $11-4+1 = 8$. Despite different widths of these 9 channels, the max-over-time pooling gives a concatenated 9-dimensional vector, which is finally transformed into a 2-dimensional output vector for binary sentiment predictions.

## Defining the Model

We implement the textCNN model in the following class. Compared with the bidirectional RNN model in Section 16.2, besides replacing recurrent layers with convolutional layers, we also use two embedding layers: one with trainable weights and the other with fixed weights.

```
class TextCNN(nn.Module):
    def __init__(self, vocab_size, embed_size, kernel_sizes, num_channels,
                 **kwargs):
        super(TextCNN, self).__init__(**kwargs)
        self.embedding = nn.Embedding(vocab_size, embed_size)
        # The embedding layer not to be trained
        self.constant_embedding = nn.Embedding(vocab_size, embed_size)
        self.dropout = nn.Dropout(0.5)
        self.decoder = nn.Linear(sum(num_channels), 2)
        # The max-over-time pooling layer has no parameters, so this instance
        # can be shared
        self.pool = nn.AdaptiveAvgPool1d(1)
        self.relu = nn.ReLU()
        # Create multiple one-dimensional convolutional layers
        self.convs = nn.ModuleList()
        for c, k in zip(num_channels, kernel_sizes):
            self.convs.append(nn.Conv1d(2 * embed_size, c, k))

    def forward(self, inputs):
        # Concatenate two embedding layer outputs with shape (batch size, no.
        # of tokens, token vector dimension) along vectors
        embeddings = torch.cat((
            self.embedding(inputs), self.constant_embedding(inputs)), dim=2)
        # Per the input format of one-dimensional convolutional layers,
        # rearrange the tensor so that the second dimension stores channels
        embeddings = embeddings.permute(0, 2, 1)
        # For each one-dimensional convolutional layer, after max-over-time
        # pooling, a tensor of shape (batch size, no. of channels, 1) is
        # obtained. Remove the last dimension and concatenate along channels
        encoding = torch.cat([
            torch.squeeze(self.relu(self.pool(conv(embeddings))), dim=-1)
            for conv in self.convs], dim=1)
        outputs = self.decoder(self.dropout(encoding))
        return outputs
```

Let's create a textCNN instance. It has 3 convolutional layers with kernel widths of 3, 4, and 5, all with 100 output channels.

```
embed_size, kernel_sizes, nums_channels = 100, [3, 4, 5], [100, 100, 100]
devices = d2l.try_all_gpus()
net = TextCNN(len(vocab), embed_size, kernel_sizes, nums_channels)

def init_weights(module):
    if type(module) in (nn.Linear, nn.Conv1d):
        nn.init.xavier_uniform_(module.weight)

net.apply(init_weights);
```

### Loading Pretrained Word Vectors

Same as Section 16.2, we load pretrained 100-dimensional GloVe embeddings as the initialized token representations. These token representations (embedding weights) will be trained in `embedding` and fixed in `constant_embedding`.

```
glove_embedding = d2l.TokenEmbedding('glove.6b.100d')
embeds = glove_embedding[vocab.idx_to_token]
net.embedding.weight.data.copy_(embeds)
net.constant_embedding.weight.data.copy_(embeds)
net.constant_embedding.weight.requires_grad = False
```

### Training and Evaluating the Model

Now we can train the textCNN model for sentiment analysis.

```
lr, num_epochs = 0.001, 5
trainer = torch.optim.Adam(net.parameters(), lr=lr)
loss = nn.CrossEntropyLoss(reduction="none")
d2l.train_ch13(net, train_iter, test_iter, loss, trainer, num_epochs, devices)
```

```
loss 0.066, train acc 0.979, test acc 0.871
2832.0 examples/sec on [device(type='cuda', index=0), device(type='cuda',␣
↪index=1)]
```

Below we use the trained model to predict the sentiment for two simple sentences.

```
d2l.predict_sentiment(net, vocab, 'this movie is so great')
```

```
'positive'
```

```
d2l.predict_sentiment(net, vocab, 'this movie is so bad')
```

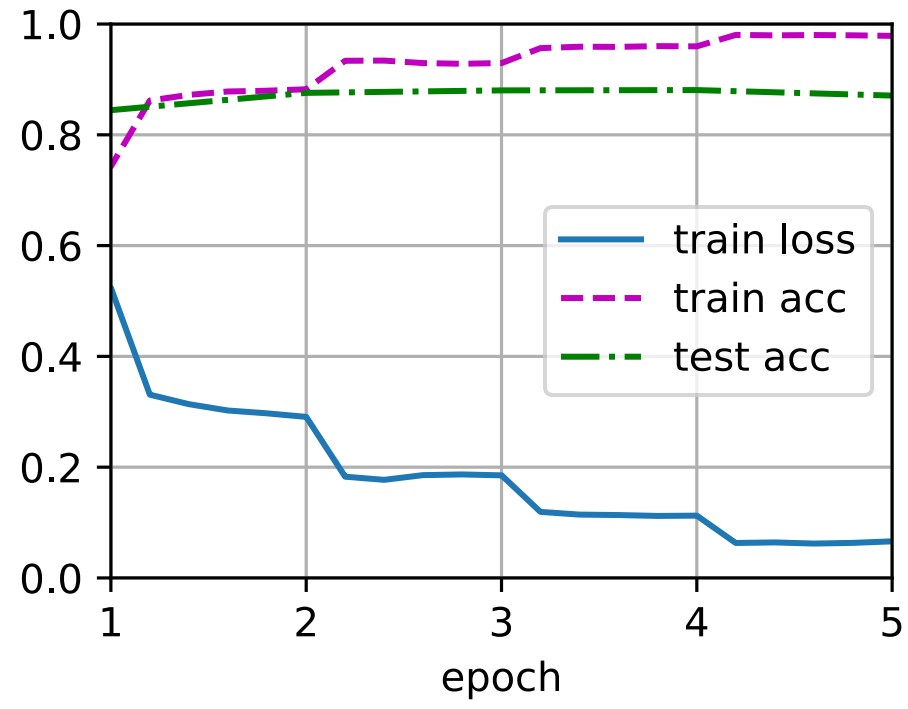

```
'negative'
```

## 16.3.4 Summary

- One-dimensional CNNs can process local features such as $n$-grams in text.
- Multi-input-channel one-dimensional cross-correlations are equivalent to single-input-channel two-dimensional cross-correlations.
- The max-over-time pooling allows different numbers of time steps at different channels.
- The textCNN model transforms individual token representations into downstream application outputs using one-dimensional convolutional layers and max-over-time pooling layers.

## 16.3.5 Exercises

1. Tune hyperparameters and compare the two architectures for sentiment analysis in Section 16.2 and in this section, such as in classification accuracy and computational efficiency.
2. Can you further improve the classification accuracy of the model by using the methods introduced in the exercises of Section 16.2?
3. Add positional encoding in the input representations. Does it improve the classification accuracy?

Discussions[245].

# 16.4 Natural Language Inference and the Dataset

In Section 16.1, we discussed the problem of sentiment analysis. This task aims to classify a single text sequence into predefined categories, such as a set of sentiment polarities. However, when there is a need to decide whether one sentence can be inferred form another, or eliminate redundancy by identifying sentences that are semantically equivalent, knowing how to classify one text sequence is insufficient. Instead, we need to be able to reason over pairs of text sequences.

## 16.4.1 Natural Language Inference

*Natural language inference* studies whether a *hypothesis* can be inferred from a *premise*, where both are a text sequence. In other words, natural language inference determines the logical relationship between a pair of text sequences. Such relationships usually fall into three types:

- *Entailment*: the hypothesis can be inferred from the premise.
- *Contradiction*: the negation of the hypothesis can be inferred from the premise.
- *Neutral*: all the other cases.

Natural language inference is also known as the recognizing textual entailment task. For example, the following pair will be labeled as *entailment* because "showing affection" in the hypothesis can be inferred from "hugging one another" in the premise.

Premise: Two women are hugging each other.

Hypothesis: Two women are showing affection.

The following is an example of *contradiction* as "running the coding example" indicates "not sleeping" rather than "sleeping".

Premise: A man is running the coding example from Dive into Deep Learning.

Hypothesis: The man is sleeping.

The third example shows a *neutrality* relationship because neither "famous" nor "not famous" can be inferred from the fact that "are performing for us".

Premise: The musicians are performing for us.

Hypothesis: The musicians are famous.

Natural language inference has been a central topic for understanding natural language. It enjoys wide applications ranging from information retrieval to open-domain question answering. To study this problem, we will begin by investigating a popular natural language inference benchmark dataset.

## 16.4.2 The Stanford Natural Language Inference (SNLI) Dataset

Stanford Natural Language Inference (SNLI) Corpus is a collection of over 500000 labeled English sentence pairs (Bowman *et al.*, 2015). We download and store the extracted SNLI dataset in the path `../data/snli_1.0`.

```
import os
import re
import torch
from torch import nn
from d2l import torch as d2l

#@save
d2l.DATA_HUB['SNLI'] = (
    'https://nlp.stanford.edu/projects/snli/snli_1.0.zip',
    '9fcde07509c7e87ec61c640c1b2753d9041758e4')

data_dir = d2l.download_extract('SNLI')
```

### Reading the Dataset

The original SNLI dataset contains much richer information than what we really need in our experiments. Thus, we define a function `read_snli` to only extract part of the dataset, then return lists of premises, hypotheses, and their labels.

```
#@save
def read_snli(data_dir, is_train):
    """Read the SNLI dataset into premises, hypotheses, and labels."""
    def extract_text(s):
        # Remove information that will not be used by us
        s = re.sub('\\(', '', s)
        s = re.sub('\\)', '', s)
        # Substitute two or more consecutive whitespace with space
        s = re.sub('\\s{2,}', ' ', s)
        return s.strip()
    label_set = {'entailment': 0, 'contradiction': 1, 'neutral': 2}
    file_name = os.path.join(data_dir, 'snli_1.0_train.txt'
                             if is_train else 'snli_1.0_test.txt')
    with open(file_name, 'r') as f:
        rows = [row.split('\t') for row in f.readlines()[1:]]
    premises = [extract_text(row[1]) for row in rows if row[0] in label_set]
    hypotheses = [extract_text(row[2]) for row in rows if row[0] in label_set]
    labels = [label_set[row[0]] for row in rows if row[0] in label_set]
    return premises, hypotheses, labels
```

Now let's print the first 3 pairs of premise and hypothesis, as well as their labels ("0", "1", and "2" correspond to "entailment", "contradiction", and "neutral", respectively ).

```
train_data = read_snli(data_dir, is_train=True)
```

```
for x0, x1, y in zip(train_data[0][:3], train_data[1][:3], train_data[2][:3]):
    print('premise:', x0)
    print('hypothesis:', x1)
    print('label:', y)
```

```
premise: A person on a horse jumps over a broken down airplane .
hypothesis: A person is training his horse for a competition .
label: 2
premise: A person on a horse jumps over a broken down airplane .
hypothesis: A person is at a diner , ordering an omelette .
label: 1
premise: A person on a horse jumps over a broken down airplane .
hypothesis: A person is outdoors , on a horse .
label: 0
```

The training set has about 550000 pairs, and the testing set has about 10000 pairs. The following shows that the three labels "entailment", "contradiction", and "neutral" are balanced in both the training set and the testing set.

```
test_data = read_snli(data_dir, is_train=False)
for data in [train_data, test_data]:
    print([[row for row in data[2]].count(i) for i in range(3)])
```

```
[183416, 183187, 182764]
[3368, 3237, 3219]
```

### Defining a Class for Loading the Dataset

Below we define a class for loading the SNLI dataset by inheriting from the `Dataset` class in Gluon. The argument `num_steps` in the class constructor specifies the length of a text sequence so that each minibatch of sequences will have the same shape. In other words, tokens after the first `num_steps` ones in longer sequence are trimmed, while special tokens "<pad>" will be appended to shorter sequences until their length becomes `num_steps`. By implementing the `__getitem__` function, we can arbitrarily access the premise, hypothesis, and label with the index `idx`.

```
#@save
class SNLIDataset(torch.utils.data.Dataset):
    """A customized dataset to load the SNLI dataset."""
    def __init__(self, dataset, num_steps, vocab=None):
        self.num_steps = num_steps
        all_premise_tokens = d2l.tokenize(dataset[0])
        all_hypothesis_tokens = d2l.tokenize(dataset[1])
        if vocab is None:
            self.vocab = d2l.Vocab(all_premise_tokens + all_hypothesis_tokens,
```

```
                                    min_freq=5, reserved_tokens=['<pad>'])
        else:
            self.vocab = vocab
        self.premises = self._pad(all_premise_tokens)
        self.hypotheses = self._pad(all_hypothesis_tokens)
        self.labels = torch.tensor(dataset[2])
        print('read ' + str(len(self.premises)) + ' examples')

    def _pad(self, lines):
        return torch.tensor([d2l.truncate_pad(
            self.vocab[line], self.num_steps, self.vocab['<pad>'])
                         for line in lines])

    def __getitem__(self, idx):
        return (self.premises[idx], self.hypotheses[idx]), self.labels[idx]

    def __len__(self):
        return len(self.premises)
```

### Putting It All Together

Now we can invoke the `read_snli` function and the `SNLIDataset` class to download the SNLI dataset and return `DataLoader` instances for both training and testing sets, together with the vocabulary of the training set. It is noteworthy that we must use the vocabulary constructed from the training set as that of the testing set. As a result, any new token from the testing set will be unknown to the model trained on the training set.

```
#@save
def load_data_snli(batch_size, num_steps=50):
    """Download the SNLI dataset and return data iterators and vocabulary."""
    num_workers = d2l.get_dataloader_workers()
    data_dir = d2l.download_extract('SNLI')
    train_data = read_snli(data_dir, True)
    test_data = read_snli(data_dir, False)
    train_set = SNLIDataset(train_data, num_steps)
    test_set = SNLIDataset(test_data, num_steps, train_set.vocab)
    train_iter = torch.utils.data.DataLoader(train_set, batch_size,
                                             shuffle=True,
                                             num_workers=num_workers)
    test_iter = torch.utils.data.DataLoader(test_set, batch_size,
                                            shuffle=False,
                                            num_workers=num_workers)
    return train_iter, test_iter, train_set.vocab
```

Here we set the batch size to 128 and sequence length to 50, and invoke the `load_data_snli` function to get the data iterators and vocabulary. Then we print the vocabulary size.

```
train_iter, test_iter, vocab = load_data_snli(128, 50)
len(vocab)
```

```
read 549367 examples
read 9824 examples
```

```
18678
```

Now we print the shape of the first minibatch. Contrary to sentiment analysis, we have two inputs `X[0]` and `X[1]` representing pairs of premises and hypotheses.

```
for X, Y in train_iter:
    print(X[0].shape)
    print(X[1].shape)
    print(Y.shape)
    break
```

```
torch.Size([128, 50])
torch.Size([128, 50])
torch.Size([128])
```

### 16.4.3 Summary

- Natural language inference studies whether a hypothesis can be inferred from a premise, where both are a text sequence.
- In natural language inference, relationships between premises and hypotheses include entailment, contradiction, and neutral.
- Stanford Natural Language Inference (SNLI) Corpus is a popular benchmark dataset of natural language inference.

### 16.4.4 Exercises

1. Machine translation has long been evaluated based on superficial $n$-gram matching between an output translation and a ground-truth translation. Can you design a measure for evaluating machine translation results by using natural language inference?
2. How can we change hyperparameters to reduce the vocabulary size?

Discussions[246].

# 16.5 Natural Language Inference: Using Attention

We introduced the natural language inference task and the SNLI dataset in Section 16.4. In view of many models that are based on complex and deep architectures, Parikh *et al.* (2016) proposed to address natural language inference with attention mechanisms and called it a "decomposable attention model". This results in a model without recurrent or convolutional layers, achieving the best result at the time on the SNLI dataset with much fewer parameters. In this section, we will describe and implement this attention-based method (with MLPs) for natural language inference, as depicted in Fig. 16.5.1.

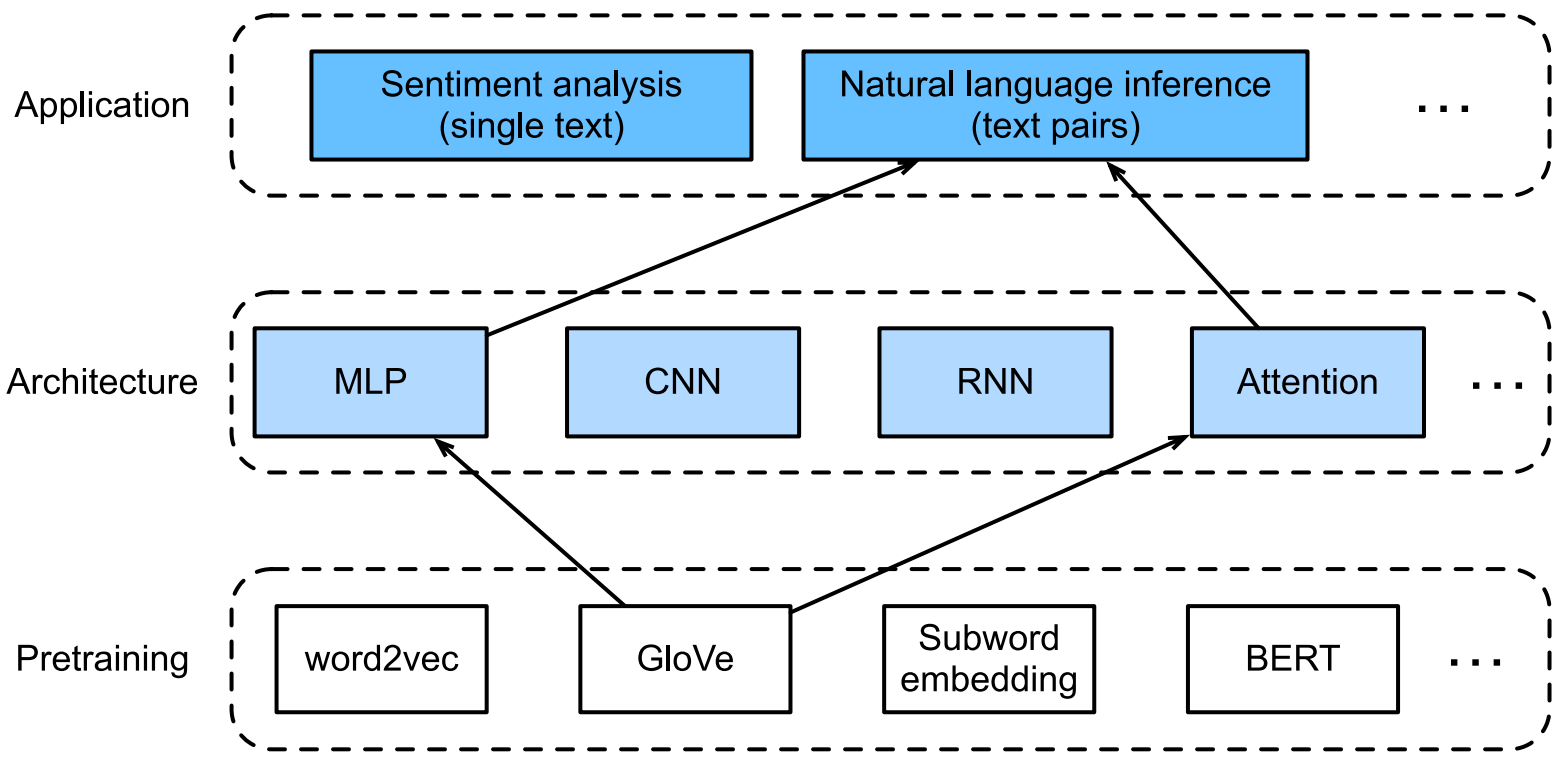

Fig. 16.5.1 This section feeds pretrained GloVe to an architecture based on attention and MLPs for natural language inference.

## 16.5.1 The Model

Simpler than preserving the order of tokens in premises and hypotheses, we can just align tokens in one text sequence to every token in the other, and vice versa, then compare and aggregate such information to predict the logical relationships between premises and hypotheses. Similar to alignment of tokens between source and target sentences in machine translation, the alignment of tokens between premises and hypotheses can be neatly accomplished by attention mechanisms.

Fig. 16.5.2 depicts the natural language inference method using attention mechanisms. At a high level, it consists of three jointly trained steps: attending, comparing, and aggregating. We will illustrate them step by step in the following.

```
import torch
from torch import nn
from torch.nn import functional as F
from d2l import torch as d2l
```

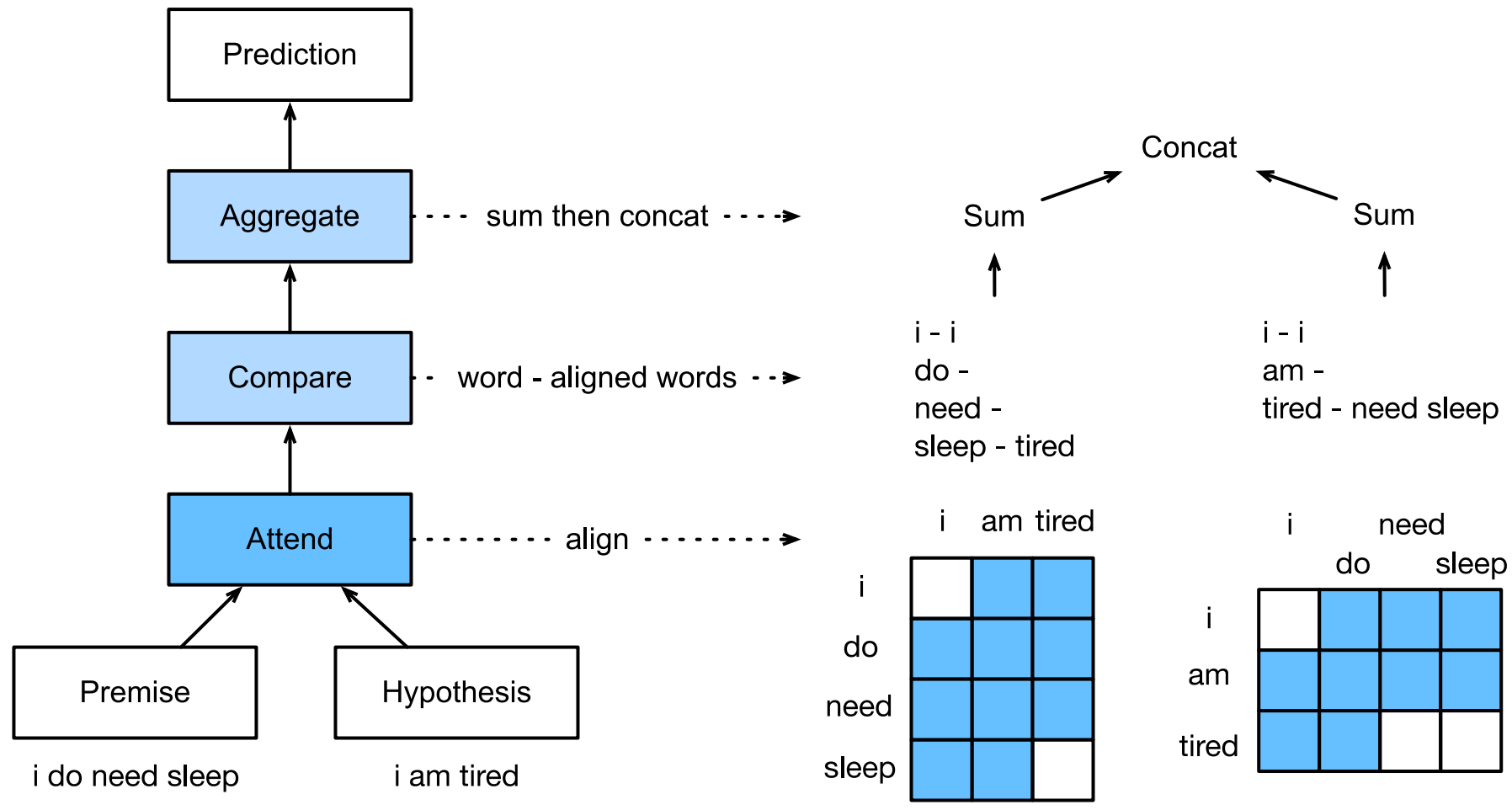

Fig. 16.5.2 Natural language inference using attention mechanisms.

### Attending

The first step is to align tokens in one text sequence to each token in the other sequence. Suppose that the premise is "i do need sleep" and the hypothesis is "i am tired". Due to semantical similarity, we may wish to align "i" in the hypothesis with "i" in the premise, and align "tired" in the hypothesis with "sleep" in the premise. Likewise, we may wish to align "i" in the premise with "i" in the hypothesis, and align "need" and "sleep" in the premise with "tired" in the hypothesis. Note that such alignment is *soft* using weighted average, where ideally large weights are associated with the tokens to be aligned. For ease of demonstration, Fig. 16.5.2 shows such alignment in a *hard* way.

Now we describe the soft alignment using attention mechanisms in more detail. Denote by $\mathbf{A} = (\mathbf{a}_1, \ldots, \mathbf{a}_m)$ and $\mathbf{B} = (\mathbf{b}_1, \ldots, \mathbf{b}_n)$ the premise and hypothesis, whose number of tokens are $m$ and $n$, respectively, where $\mathbf{a}_i, \mathbf{b}_j \in \mathbb{R}^d$ $(i = 1, \ldots, m, j = 1, \ldots, n)$ is a $d$-dimensional word vector. For soft alignment, we compute the attention weights $e_{ij} \in \mathbb{R}$ as

$$e_{ij} = f(\mathbf{a}_i)^\top f(\mathbf{b}_j), \tag{16.5.1}$$

where the function $f$ is an MLP defined in the following `mlp` function. The output dimension of $f$ is specified by the `num_hiddens` argument of `mlp`.

```
def mlp(num_inputs, num_hiddens, flatten):
    net = []
    net.append(nn.Dropout(0.2))
    net.append(nn.Linear(num_inputs, num_hiddens))
    net.append(nn.ReLU())
    if flatten:
        net.append(nn.Flatten(start_dim=1))
    net.append(nn.Dropout(0.2))
```

```
    net.append(nn.Linear(num_hiddens, num_hiddens))
    net.append(nn.ReLU())
    if flatten:
        net.append(nn.Flatten(start_dim=1))
    return nn.Sequential(*net)
```

It should be highlighted that, in (16.5.1) $f$ takes inputs $\mathbf{a}_i$ and $\mathbf{b}_j$ separately rather than takes a pair of them together as input. This *decomposition* trick leads to only $m + n$ applications (linear complexity) of $f$ rather than $mn$ applications (quadratic complexity).

Normalizing the attention weights in (16.5.1), we compute the weighted average of all the token vectors in the hypothesis to obtain representation of the hypothesis that is softly aligned with the token indexed by $i$ in the premise:

$$\boldsymbol{\beta}_i = \sum_{j=1}^{n}\frac{\exp(e_{ij})}{ \sum_{k=1}^{n} \exp(e_{ik})} \mathbf{b}_j. \tag{16.5.2}$$

Likewise, we compute soft alignment of premise tokens for each token indexed by $j$ in the hypothesis:

$$\boldsymbol{\alpha}_j = \sum_{i=1}^{m}\frac{\exp(e_{ij})}{ \sum_{k=1}^{m} \exp(e_{kj})} \mathbf{a}_i. \tag{16.5.3}$$

Below we define the `Attend` class to compute the soft alignment of hypotheses (`beta`) with input premises `A` and soft alignment of premises (`alpha`) with input hypotheses `B`.

```
class Attend(nn.Module):
    def __init__(self, num_inputs, num_hiddens, **kwargs):
        super(Attend, self).__init__(**kwargs)
        self.f = mlp(num_inputs, num_hiddens, flatten=False)

    def forward(self, A, B):
        # Shape of `A`/`B`: (`batch_size`, no. of tokens in sequence A/B,
        # `embed_size`)
        # Shape of `f_A`/`f_B`: (`batch_size`, no. of tokens in sequence A/B,
        # `num_hiddens`)
        f_A = self.f(A)
        f_B = self.f(B)
        # Shape of `e`: (`batch_size`, no. of tokens in sequence A,
        # no. of tokens in sequence B)
        e = torch.bmm(f_A, f_B.permute(0, 2, 1))
        # Shape of `beta`: (`batch_size`, no. of tokens in sequence A,
        # `embed_size`), where sequence B is softly aligned with each token
        # (axis 1 of `beta`) in sequence A
        beta = torch.bmm(F.softmax(e, dim=-1), B)
        # Shape of `alpha`: (`batch_size`, no. of tokens in sequence B,
        # `embed_size`), where sequence A is softly aligned with each token
        # (axis 1 of `alpha`) in sequence B
        alpha = torch.bmm(F.softmax(e.permute(0, 2, 1), dim=-1), A)
        return beta, alpha
```

### Comparing

In the next step, we compare a token in one sequence with the other sequence that is softly aligned with that token. Note that in soft alignment, all the tokens from one sequence, though with probably different attention weights, will be compared with a token in the other sequence. For easy of demonstration, Fig. 16.5.2 pairs tokens with aligned tokens in a *hard* way. For example, suppose that the attending step determines that "need" and "sleep" in the premise are both aligned with "tired" in the hypothesis, the pair "tired–need sleep" will be compared.

In the comparing step, we feed the concatenation (operator $[\cdot, \cdot]$) of tokens from one sequence and aligned tokens from the other sequence into a function $g$ (an MLP):

$$\begin{aligned} \mathbf{v}_{A,i} &= g([\mathbf{a}_i, \boldsymbol{\beta}_i]), i = 1, \ldots, m \\ \mathbf{v}_{B,j} &= g([\mathbf{b}_j, \boldsymbol{\alpha}_j]), j = 1, \ldots, n. \end{aligned} \tag{16.5.4}$$

In (16.5.4), $\mathbf{v}_{A,i}$ is the comparison between token $i$ in the premise and all the hypothesis tokens that are softly aligned with token $i$; while $\mathbf{v}_{B,j}$ is the comparison between token $j$ in the hypothesis and all the premise tokens that are softly aligned with token $j$. The following `Compare` class defines such as comparing step.

```
class Compare(nn.Module):
    def __init__(self, num_inputs, num_hiddens, **kwargs):
        super(Compare, self).__init__(**kwargs)
        self.g = mlp(num_inputs, num_hiddens, flatten=False)

    def forward(self, A, B, beta, alpha):
        V_A = self.g(torch.cat([A, beta], dim=2))
        V_B = self.g(torch.cat([B, alpha], dim=2))
        return V_A, V_B
```

### Aggregating

With two sets of comparison vectors $\mathbf{v}_{A,i}$ ($i = 1, \ldots, m$) and $\mathbf{v}_{B,j}$ ($j = 1, \ldots, n$) on hand, in the last step we will aggregate such information to infer the logical relationship. We begin by summing up both sets:

$$\mathbf{v}_A = \sum_{i=1}^{m} \mathbf{v}_{A,i}, \quad \mathbf{v}_B = \sum_{j=1}^{n} \mathbf{v}_{B,j}. \tag{16.5.5}$$

Next we feed the concatenation of both summarization results into function $h$ (an MLP) to obtain the classification result of the logical relationship:

$$\hat{\mathbf{y}} = h([\mathbf{v}_A, \mathbf{v}_B]). \tag{16.5.6}$$

The aggregation step is defined in the following `Aggregate` class.

```
class Aggregate(nn.Module):
    def __init__(self, num_inputs, num_hiddens, num_outputs, **kwargs):
        super(Aggregate, self).__init__(**kwargs)
        self.h = mlp(num_inputs, num_hiddens, flatten=True)
        self.linear = nn.Linear(num_hiddens, num_outputs)

    def forward(self, V_A, V_B):
        # Sum up both sets of comparison vectors
        V_A = V_A.sum(dim=1)
        V_B = V_B.sum(dim=1)
        # Feed the concatenation of both summarization results into an MLP
        Y_hat = self.linear(self.h(torch.cat([V_A, V_B], dim=1)))
        return Y_hat
```

### Putting It All Together

By putting the attending, comparing, and aggregating steps together, we define the decomposable attention model to jointly train these three steps.

```
class DecomposableAttention(nn.Module):
    def __init__(self, vocab, embed_size, num_hiddens, num_inputs_attend=100,
                 num_inputs_compare=200, num_inputs_agg=400, **kwargs):
        super(DecomposableAttention, self).__init__(**kwargs)
        self.embedding = nn.Embedding(len(vocab), embed_size)
        self.attend = Attend(num_inputs_attend, num_hiddens)
        self.compare = Compare(num_inputs_compare, num_hiddens)
        # There are 3 possible outputs: entailment, contradiction, and neutral
        self.aggregate = Aggregate(num_inputs_agg, num_hiddens, num_outputs=3)

    def forward(self, X):
        premises, hypotheses = X
        A = self.embedding(premises)
        B = self.embedding(hypotheses)
        beta, alpha = self.attend(A, B)
        V_A, V_B = self.compare(A, B, beta, alpha)
        Y_hat = self.aggregate(V_A, V_B)
        return Y_hat
```

## 16.5.2 Training and Evaluating the Model

Now we will train and evaluate the defined decomposable attention model on the SNLI dataset. We begin by reading the dataset.

### Reading the dataset

We download and read the SNLI dataset using the function defined in Section 16.4. The batch size and sequence length are set to $256$ and $50$, respectively.

```
batch_size, num_steps = 256, 50
train_iter, test_iter, vocab = d2l.load_data_snli(batch_size, num_steps)
```

```
read 549367 examples
read 9824 examples
```

### Creating the Model

We use the pretrained 100-dimensional GloVe embedding to represent the input tokens. Thus, we predefine the dimension of vectors $\mathbf{a}_i$ and $\mathbf{b}_j$ in (16.5.1) as 100. The output dimension of functions $f$ in (16.5.1) and $g$ in (16.5.4) is set to 200. Then we create a model instance, initialize its parameters, and load the GloVe embedding to initialize vectors of input tokens.

```
embed_size, num_hiddens, devices = 100, 200, d2l.try_all_gpus()
net = DecomposableAttention(vocab, embed_size, num_hiddens)
glove_embedding = d2l.TokenEmbedding('glove.6b.100d')
embeds = glove_embedding[vocab.idx_to_token]
net.embedding.weight.data.copy_(embeds);
```

### Training and Evaluating the Model

In contrast to the `split_batch` function in Section 13.5 that takes single inputs such as text sequences (or images), we define a `split_batch_multi_inputs` function to take multiple inputs such as premises and hypotheses in minibatches.

Now we can train and evaluate the model on the SNLI dataset.

```
lr, num_epochs = 0.001, 4
trainer = torch.optim.Adam(net.parameters(), lr=lr)
loss = nn.CrossEntropyLoss(reduction="none")
d2l.train_ch13(net, train_iter, test_iter, loss, trainer, num_epochs, devices)
```

```
loss 0.496, train acc 0.805, test acc 0.823
9660.5 examples/sec on [device(type='cuda', index=0), device(type='cuda',␣
↪index=1)]
```

### Using the Model

Finally, define the prediction function to output the logical relationship between a pair of premise and hypothesis.

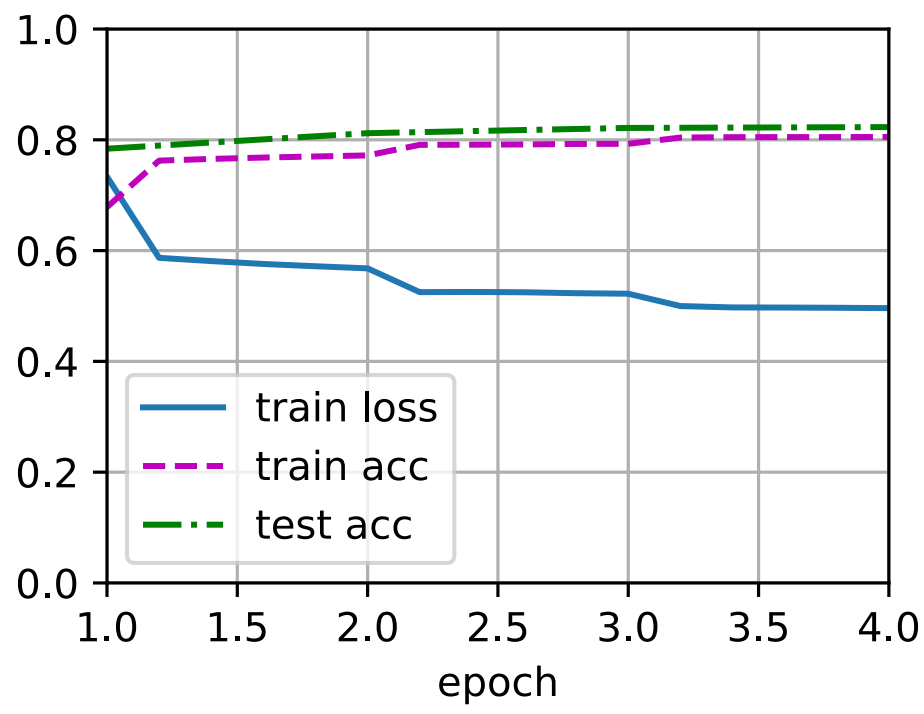

```
#@save
def predict_snli(net, vocab, premise, hypothesis):
    """Predict the logical relationship between the premise and hypothesis."""
    net.eval()
    premise = torch.tensor(vocab[premise], device=d2l.try_gpu())
    hypothesis = torch.tensor(vocab[hypothesis], device=d2l.try_gpu())
    label = torch.argmax(net([premise.reshape((1, -1)),
                              hypothesis.reshape((1, -1))]), dim=1)
    return 'entailment' if label == 0 else 'contradiction' if label == 1 \
            else 'neutral'
```

We can use the trained model to obtain the natural language inference result for a sample pair of sentences.

```
predict_snli(net, vocab, ['he', 'is', 'good', '.'], ['he', 'is', 'bad', '.'])
```

```
'contradiction'
```

### 16.5.3 Summary

- The decomposable attention model consists of three steps for predicting the logical relationships between premises and hypotheses: attending, comparing, and aggregating.
- With attention mechanisms, we can align tokens in one text sequence to every token in the other, and vice versa. Such alignment is soft using weighted average, where ideally large weights are associated with the tokens to be aligned.
- The decomposition trick leads to a more desirable linear complexity than quadratic complexity when computing attention weights.
- We can use pretrained word vectors as the input representation for downstream natural language processing task such as natural language inference.

### 16.5.4 Exercises

1. Train the model with other combinations of hyperparameters. Can you get better accuracy on the test set?
2. What are major drawbacks of the decomposable attention model for natural language inference?
3. Suppose that we want to get the level of semantical similarity (e.g., a continuous value between 0 and 1) for any pair of sentences. How shall we collect and label the dataset? Can you design a model with attention mechanisms?

Discussions[247].

## 16.6 Fine-Tuning BERT for Sequence-Level and Token-Level Applications

In the previous sections of this chapter, we have designed different models for natural language processing applications, such as based on RNNs, CNNs, attention, and MLPs. These models are helpful when there is space or time constraint, however, crafting a specific model for every natural language processing task is practically infeasible. In Section 15.8, we introduced a pretraining model, BERT, that requires minimal architecture changes for a wide range of natural language processing tasks. On the one hand, at the time of its proposal, BERT improved the state of the art on various natural language processing tasks. On the other hand, as noted in Section 15.10, the two versions of the original BERT model come with 110 million and 340 million parameters. Thus, when there are sufficient computational resources, we may consider fine-tuning BERT for downstream natural language processing applications.

In the following, we generalize a subset of natural language processing applications as sequence-level and token-level. On the sequence level, we introduce how to transform the BERT representation of the text input to the output label in single text classification and text pair classification or regression. On the token level, we will briefly introduce new applications such as text tagging and question answering and shed light on how BERT can represent their inputs and get transformed into output labels. During fine-tuning, the "minimal architecture changes" required by BERT across different applications are the extra fully connected layers. During supervised learning of a downstream application, parameters of the extra layers are learned from scratch while all the parameters in the pretrained BERT model are fine-tuned.

### 16.6.1 Single Text Classification

*Single text classification* takes a single text sequence as input and outputs its classification result. Besides sentiment analysis that we have studied in this chapter, the Corpus of Linguistic Acceptability (CoLA) is also a dataset for single text classification, judging whether a given sentence is grammatically acceptable or not (Warstadt *et al.*, 2019). For instance, “I should study.” is acceptable but “I should studying.” is not.

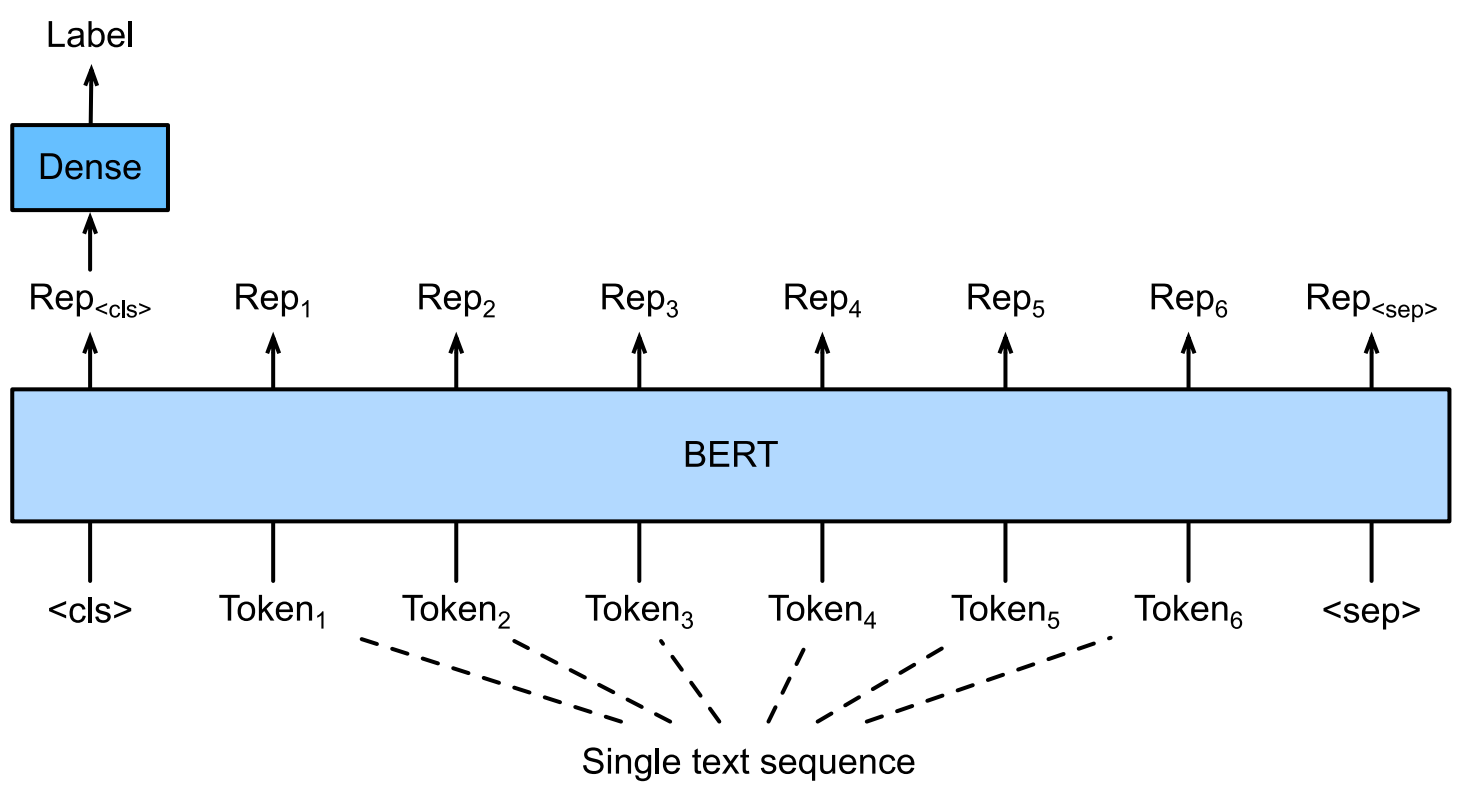

Fig. 16.6.1 Fine-tuning BERT for single text classification applications, such as sentiment analysis and testing linguistic acceptability. Suppose that the input single text has six tokens.

Section 15.8 describes the input representation of BERT. The BERT input sequence unambiguously represents both single text and text pairs, where the special classification token “<cls>” is used for sequence classification and the special classification token “<sep>” marks the end of single text or separates a pair of text. As shown in Fig. 16.6.1, in single text classification applications, the BERT representation of the special classification token “<cls>” encodes the information of the entire input text sequence. As the representation of the input single text, it will be fed into a small MLP consisting of fully connected (dense) layers to output the distribution of all the discrete label values.

## 16.6.2 Text Pair Classification or Regression

We have also examined natural language inference in this chapter. It belongs to *text pair classification*, a type of application classifying a pair of text.

Taking a pair of text as input but outputting a continuous value, *semantic textual similarity* is a popular *text pair regression* task. This task measures semantic similarity of sentences. For instance, in the Semantic Textual Similarity Benchmark dataset, the similarity score of a pair of sentences is an ordinal scale ranging from 0 (no meaning overlap) to 5 (meaning equivalence) (Cer *et al.*, 2017). The goal is to predict these scores. Examples from the Semantic Textual Similarity Benchmark dataset include (sentence 1, sentence 2, similarity score):

- “A plane is taking off.”, “An air plane is taking off.”, 5.000;
- “A woman is eating something.”, “A woman is eating meat.”, 3.000;

- “A woman is dancing.”, “A man is talking.”, 0.000.

Fig. 16.6.2 Fine-tuning BERT for text pair classification or regression applications, such as natural language inference and semantic textual similarity. Suppose that the input text pair has two and three tokens.

Comparing with single text classification in Fig. 16.6.1, fine-tuning BERT for text pair classification in Fig. 16.6.2 is different in the input representation. For text pair regression tasks such as semantic textual similarity, trivial changes can be applied such as outputting a continuous label value and using the mean squared loss: they are common for regression.

## 16.6.3 Text Tagging

Now let’s consider token-level tasks, such as *text tagging*, where each token is assigned a label. Among text tagging tasks, *part-of-speech tagging* assigns each word a part-of-speech tag (e.g., adjective and determiner) according to the role of the word in the sentence. For example, according to the Penn Treebank II tag set, the sentence “John Smith ‘s car is new” should be tagged as “NNP (noun, proper singular) NNP POS (possessive ending) NN (noun, singular or mass) VB (verb, base form) JJ (adjective)”.

Fine-tuning BERT for text tagging applications is illustrated in Fig. 16.6.3. Comparing with Fig. 16.6.1, the only distinction lies in that in text tagging, the BERT representation of *every token* of the input text is fed into the same extra fully connected layers to output the label of the token, such as a part-of-speech tag.

## 16.6.4 Question Answering

As another token-level application, *question answering* reflects capabilities of reading comprehension. For example, the Stanford Question Answering Dataset (SQuAD v1.1) consists of reading passages and questions, where the answer to every question is just a segment of text (text span) from the passage that the question is about (Rajpurkar *et al.*, 2016). To explain, consider a passage “Some experts report that a mask’s efficacy is inconclusive. However, mask makers insist that their products, such as N95 respirator masks, can guard against the virus." and a question "Who say that N95 respirator masks can guard against the virus?". The answer should be the text span "mask makers" in the passage. Thus, the goal in SQuAD v1.1 is to predict the start and end of the text span in the passage given a pair of question and passage.

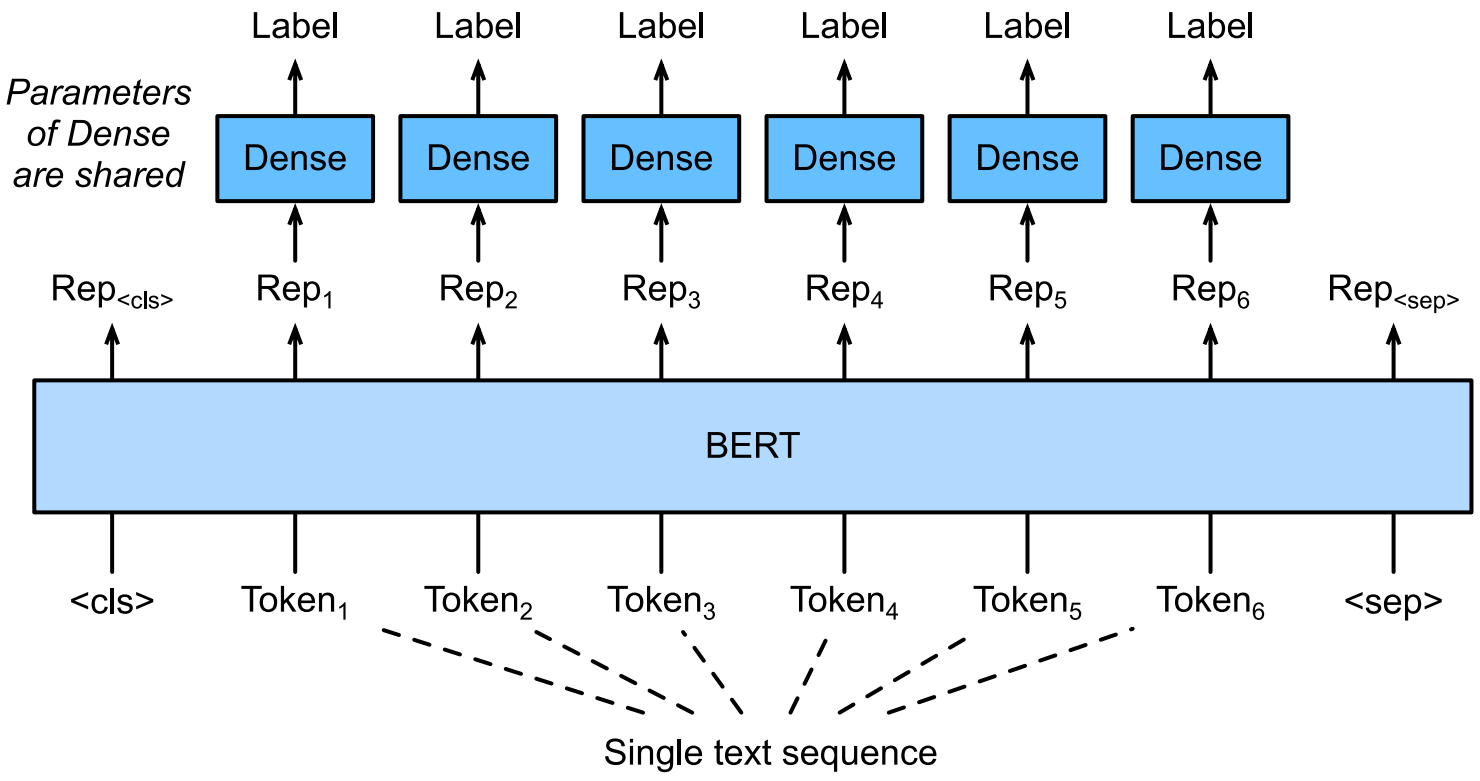

Fig. 16.6.3 Fine-tuning BERT for text tagging applications, such as part-of-speech tagging. Suppose that the input single text has six tokens.

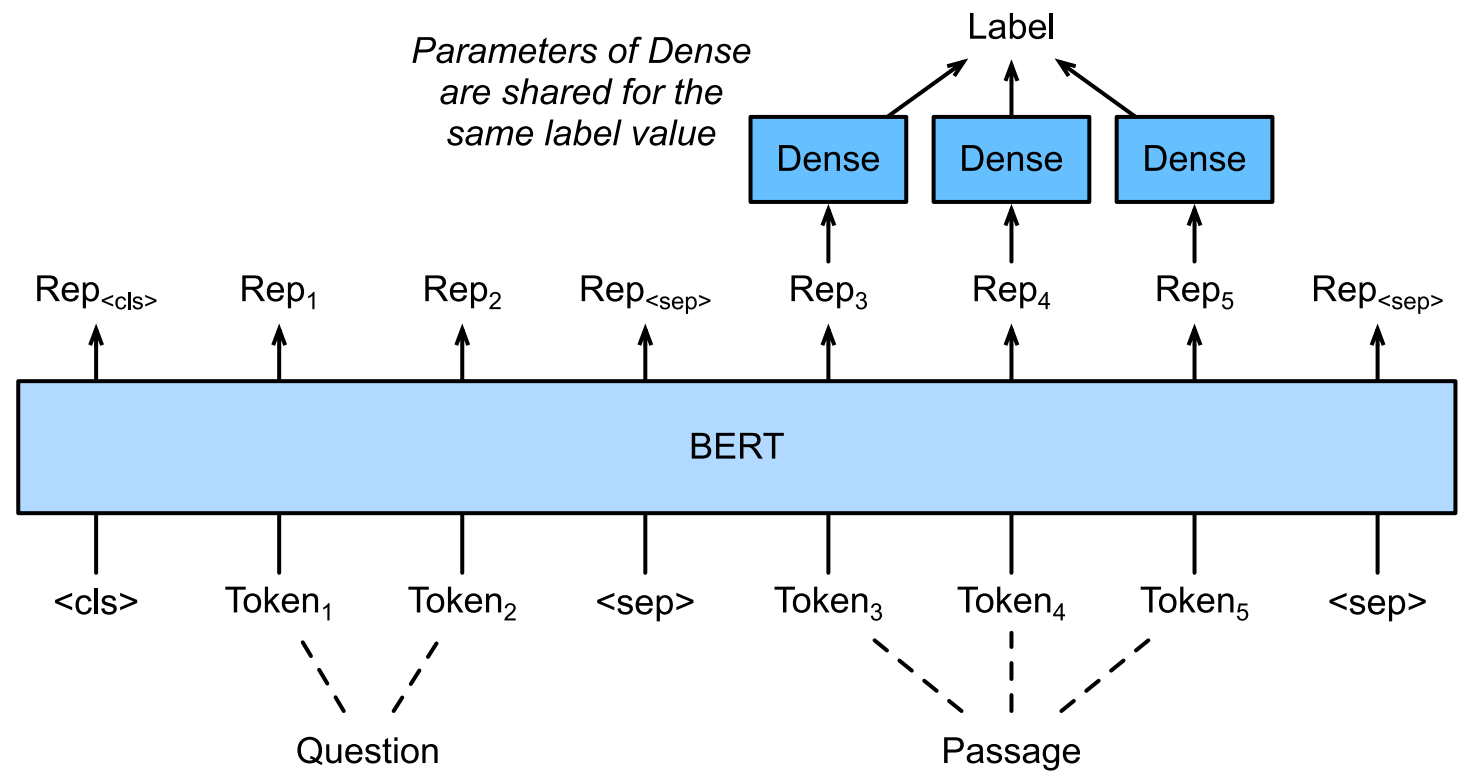

Fig. 16.6.4 Fine-tuning BERT for question answering. Suppose that the input text pair has two and three tokens.

To fine-tune BERT for question answering, the question and passage are packed as the first and second text sequence, respectively, in the input of BERT. To predict the position of the start of the text span, the same additional fully connected layer will transform the BERT representation of any token from the passage of position $i$ into a scalar score $s_i$. Such scores of all the passage tokens are further transformed by the softmax operation into a probability distribution, so that each token position $i$ in the passage is assigned a probability $p_i$ of being the start of the text span. Predicting the end of the text span is the same as above, except that parameters in its additional fully connected layer are independent from those for predicting the start. When predicting the end, any passage token of position $i$ is transformed by the same fully connected layer into a scalar score $e_i$. Fig. 16.6.4 depicts fine-tuning BERT for question answering.

For question answering, the supervised learning's training objective is as straightforward as maximizing the log-likelihoods of the ground-truth start and end positions. When predicting the span, we can compute the score $s_i + e_j$ for a valid span from position $i$ to position $j$ ($i \leq j$), and output the span with the highest score.

### 16.6.5 Summary

- BERT requires minimal architecture changes (extra fully connected layers) for sequence-level and token-level natural language processing applications, such as single text classification (e.g., sentiment analysis and testing linguistic acceptability), text pair classification or regression (e.g., natural language inference and semantic textual similarity), text tagging (e.g., part-of-speech tagging), and question answering.
- During supervised learning of a downstream application, parameters of the extra layers are learned from scratch while all the parameters in the pretrained BERT model are fine-tuned.

### 16.6.6 Exercises

1. Let's design a search engine algorithm for news articles. When the system receives an query (e.g., "oil industry during the coronavirus outbreak"), it should return a ranked list of news articles that are most relevant to the query. Suppose that we have a huge pool of news articles and a large number of queries. To simplify the problem, suppose that the most relevant article has been labeled for each query. How can we apply negative sampling (see Section 15.2.1) and BERT in the algorithm design?
2. How can we leverage BERT in training language models?
3. Can we leverage BERT in machine translation?

Discussions[248].

## 16.7 Natural Language Inference: Fine-Tuning BERT

In earlier sections of this chapter, we have designed an attention-based architecture (in Section 16.5) for the natural language inference task on the SNLI dataset (as described in Section 16.4). Now we revisit this task by fine-tuning BERT. As discussed in Section 16.6, natural language inference is a sequence-level text pair classification problem, and fine-tuning BERT only requires an additional MLP-based architecture, as illustrated in Fig. 16.7.1.

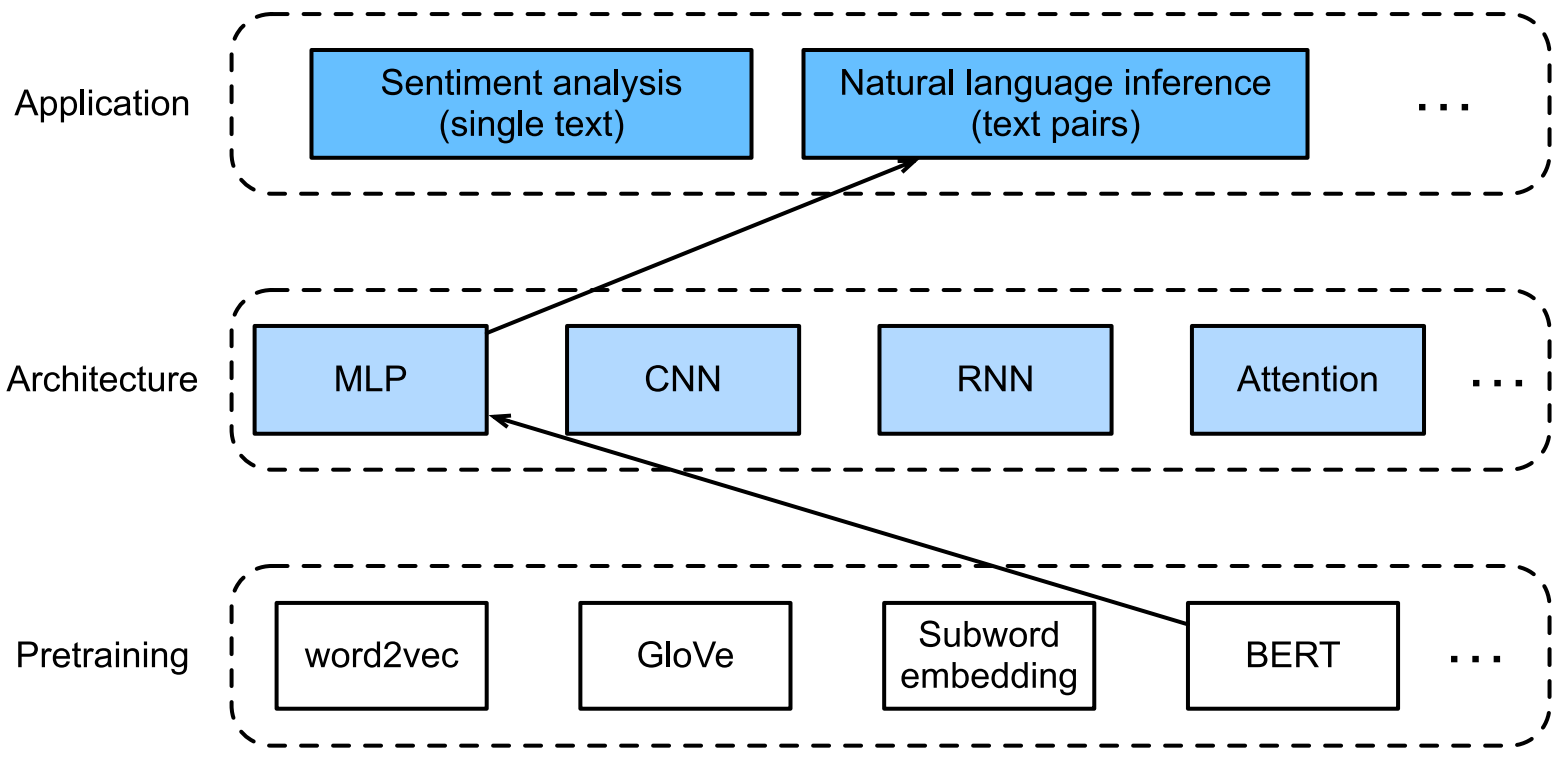

Fig. 16.7.1 This section feeds pretrained BERT to an MLP-based architecture for natural language inference.

In this section, we will download a pretrained small version of BERT, then fine-tune it for natural language inference on the SNLI dataset.

```
import json
import multiprocessing
import os
import torch
from torch import nn
from d2l import torch as d2l
```

## 16.7.1 Loading Pretrained BERT

We have explained how to pretrain BERT on the WikiText-2 dataset in Section 15.9 and Section 15.10 (note that the original BERT model is pretrained on much bigger corpora). As discussed in Section 15.10, the original BERT model has hundreds of millions of parameters. In the following, we provide two versions of pretrained BERT: "bert.base" is about as big as the original BERT base model that requires a lot of computational resources to fine-tune, while "bert.small" is a small version to facilitate demonstration.

```
d2l.DATA_HUB['bert.base'] = (d2l.DATA_URL + 'bert.base.torch.zip',
                             '225d66f04cae318b841a13d32af3acc165f253ac')
d2l.DATA_HUB['bert.small'] = (d2l.DATA_URL + 'bert.small.torch.zip',
                              'c72329e68a732bef0452e4b96a1c341c8910f81f')
```

Either pretrained BERT model contains a "vocab.json" file that defines the vocabulary set and a "pretrained.params" file of the pretrained parameters. We implement the following `load_pretrained_model` function to load pretrained BERT parameters.

```
def load_pretrained_model(pretrained_model, num_hiddens, ffn_num_hiddens,
                          num_heads, num_blks, dropout, max_len, devices):
    data_dir = d2l.download_extract(pretrained_model)
    # Define an empty vocabulary to load the predefined vocabulary
    vocab = d2l.Vocab()
    vocab.idx_to_token = json.load(open(os.path.join(data_dir, 'vocab.json')))
    vocab.token_to_idx = {token: idx for idx, token in enumerate(
        vocab.idx_to_token)}
    bert = d2l.BERTModel(
        len(vocab), num_hiddens, ffn_num_hiddens=ffn_num_hiddens, num_heads=4,
        num_blks=2, dropout=0.2, max_len=max_len)
    # Load pretrained BERT parameters
    bert.load_state_dict(torch.load(os.path.join(data_dir,
                                                 'pretrained.params')))
    return bert, vocab
```

To facilitate demonstration on most of machines, we will load and fine-tune the small version ("bert.small") of the pretrained BERT in this section. In the exercise, we will show how to fine-tune the much larger "bert.base" to significantly improve the testing accuracy.

```
devices = d2l.try_all_gpus()
bert, vocab = load_pretrained_model(
    'bert.small', num_hiddens=256, ffn_num_hiddens=512, num_heads=4,
    num_blks=2, dropout=0.1, max_len=512, devices=devices)
```

## 16.7.2 The Dataset for Fine-Tuning BERT

For the downstream task natural language inference on the SNLI dataset, we define a customized dataset class `SNLIBERTDataset`. In each example, the premise and hypothesis form a pair of text sequence and is packed into one BERT input sequence as depicted in Fig. 16.6.2. Recall Section 15.8.4 that segment IDs are used to distinguish the premise and the hypothesis in a BERT input sequence. With the predefined maximum length of a BERT input sequence (`max_len`), the last token of the longer of the input text pair keeps getting removed until `max_len` is met. To accelerate generation of the SNLI dataset for fine-tuning BERT, we use 4 worker processes to generate training or testing examples in parallel.

```
class SNLIBERTDataset(torch.utils.data.Dataset):
    def __init__(self, dataset, max_len, vocab=None):
        all_premise_hypothesis_tokens = [[
            p_tokens, h_tokens] for p_tokens, h_tokens in zip(
            *[d2l.tokenize([s.lower() for s in sentences])
              for sentences in dataset[:2]])]

        self.labels = torch.tensor(dataset[2])
        self.vocab = vocab
        self.max_len = max_len
        (self.all_token_ids, self.all_segments,
         self.valid_lens) = self._preprocess(all_premise_hypothesis_tokens)
        print('read ' + str(len(self.all_token_ids)) + ' examples')
```

```python
    def _preprocess(self, all_premise_hypothesis_tokens):
        pool = multiprocessing.Pool(4)  # Use 4 worker processes
        out = pool.map(self._mp_worker, all_premise_hypothesis_tokens)
        all_token_ids = [
            token_ids for token_ids, segments, valid_len in out]
        all_segments = [segments for token_ids, segments, valid_len in out]
        valid_lens = [valid_len for token_ids, segments, valid_len in out]
        return (torch.tensor(all_token_ids, dtype=torch.long),
                torch.tensor(all_segments, dtype=torch.long),
                torch.tensor(valid_lens))

    def _mp_worker(self, premise_hypothesis_tokens):
        p_tokens, h_tokens = premise_hypothesis_tokens
        self._truncate_pair_of_tokens(p_tokens, h_tokens)
        tokens, segments = d2l.get_tokens_and_segments(p_tokens, h_tokens)
        token_ids = self.vocab[tokens] + [self.vocab['<pad>']] \
                             * (self.max_len - len(tokens))
        segments = segments + [0] * (self.max_len - len(segments))
        valid_len = len(tokens)
        return token_ids, segments, valid_len

    def _truncate_pair_of_tokens(self, p_tokens, h_tokens):
        # Reserve slots for '<CLS>', '<SEP>', and '<SEP>' tokens for the BERT
        # input
        while len(p_tokens) + len(h_tokens) > self.max_len - 3:
            if len(p_tokens) > len(h_tokens):
                p_tokens.pop()
            else:
                h_tokens.pop()

    def __getitem__(self, idx):
        return (self.all_token_ids[idx], self.all_segments[idx],
                self.valid_lens[idx]), self.labels[idx]

    def __len__(self):
        return len(self.all_token_ids)
```

After downloading the SNLI dataset, we generate training and testing examples by instantiating the `SNLIBERTDataset` class. Such examples will be read in minibatches during training and testing of natural language inference.

```python
# Reduce `batch_size` if there is an out of memory error. In the original BERT
# model, `max_len` = 512
batch_size, max_len, num_workers = 512, 128, d2l.get_dataloader_workers()
data_dir = d2l.download_extract('SNLI')
train_set = SNLIBERTDataset(d2l.read_snli(data_dir, True), max_len, vocab)
test_set = SNLIBERTDataset(d2l.read_snli(data_dir, False), max_len, vocab)
train_iter = torch.utils.data.DataLoader(train_set, batch_size, shuffle=True,
                                   num_workers=num_workers)
test_iter = torch.utils.data.DataLoader(test_set, batch_size,
                                  num_workers=num_workers)
```

```
read 549367 examples
read 9824 examples
```

### 16.7.3 Fine-Tuning BERT

As Fig. 16.6.2 indicates, fine-tuning BERT for natural language inference requires only an extra MLP consisting of two fully connected layers (see `self.hidden` and `self.output` in the following `BERTClassifier` class). This MLP transforms the BERT representation of the special “<cls>” token, which encodes the information of both the premise and the hypothesis, into three outputs of natural language inference: entailment, contradiction, and neutral.

```
class BERTClassifier(nn.Module):
    def __init__(self, bert):
        super(BERTClassifier, self).__init__()
        self.encoder = bert.encoder
        self.hidden = bert.hidden
        self.output = nn.LazyLinear(3)

    def forward(self, inputs):
        tokens_X, segments_X, valid_lens_x = inputs
        encoded_X = self.encoder(tokens_X, segments_X, valid_lens_x)
        return self.output(self.hidden(encoded_X[:, 0, :]))
```

In the following, the pretrained BERT model `bert` is fed into the `BERTClassifier` instance `net` for the downstream application. In common implementations of BERT fine-tuning, only the parameters of the output layer of the additional MLP (`net.output`) will be learned from scratch. All the parameters of the pretrained BERT encoder (`net.encoder`) and the hidden layer of the additional MLP (`net.hidden`) will be fine-tuned.

```
net = BERTClassifier(bert)
```

Recall that in Section 15.8 both the `MaskLM` class and the `NextSentencePred` class have parameters in their employed MLPs. These parameters are part of those in the pretrained BERT model `bert`, and thus part of parameters in `net`. However, such parameters are only for computing the masked language modeling loss and the next sentence prediction loss during pretraining. These two loss functions are irrelevant to fine-tuning downstream applications, thus the parameters of the employed MLPs in `MaskLM` and `NextSentencePred` are not updated (staled) when BERT is fine-tuned.

To allow parameters with stale gradients, the flag `ignore_stale_grad=True` is set in the `step` function of `d2l.train_batch_ch13`. We use this function to train and evaluate the model `net` using the training set (`train_iter`) and the testing set (`test_iter`) of SNLI. Due to the limited computational resources, the training and testing accuracy can be further improved: we leave its discussions in the exercises.

```
lr, num_epochs = 1e-4, 5
trainer = torch.optim.Adam(net.parameters(), lr=lr)
loss = nn.CrossEntropyLoss(reduction='none')
net(next(iter(train_iter))[0])
d2l.train_ch13(net, train_iter, test_iter, loss, trainer, num_epochs, devices)
```

```
loss 0.524, train acc 0.788, test acc 0.777
8176.1 examples/sec on [device(type='cuda', index=0), device(type='cuda',␣
↪index=1)]
```

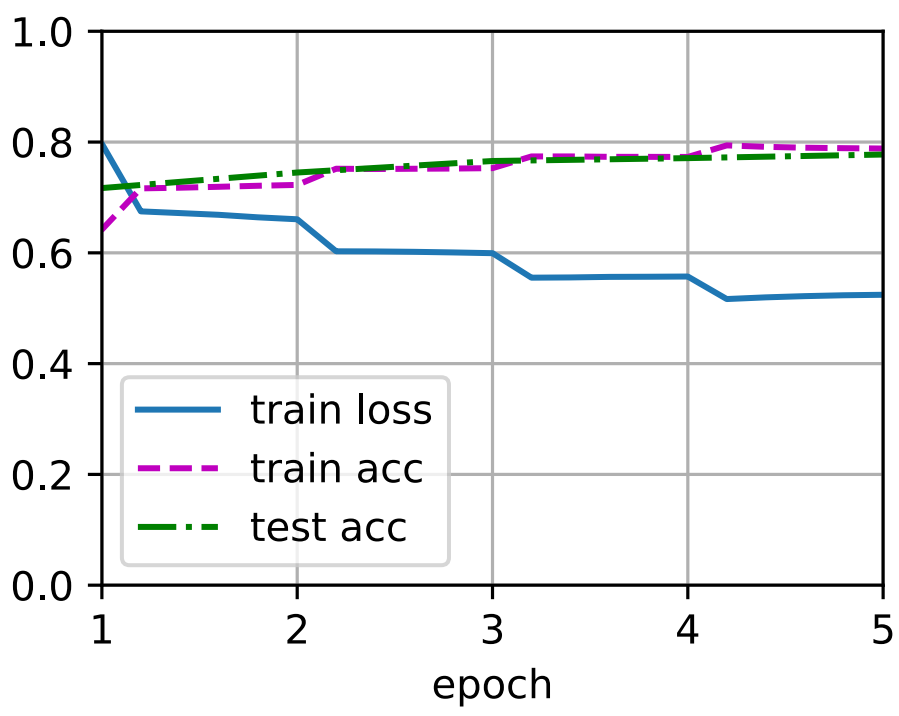

## 16.7.4 Summary

- We can fine-tune the pretrained BERT model for downstream applications, such as natural language inference on the SNLI dataset.
- During fine-tuning, the BERT model becomes part of the model for the downstream application. Parameters that are only related to pretraining loss will not be updated during fine-tuning.

## 16.7.5 Exercises

1. Fine-tune a much larger pretrained BERT model that is about as big as the original BERT base model if your computational resource allows. Set arguments in the `load_pretrained_model` function as: replacing 'bert.small' with 'bert.base', increasing values of `num_hiddens=256`, `ffn_num_hiddens=512`, `num_heads=4`, and `num_blks=2` to 768, 3072, 12, and 12, respectively. By increasing fine-tuning epochs (and possibly tuning other hyperparameters), can you get a testing accuracy higher than 0.86?
2. How to truncate a pair of sequences according to their ratio of length? Compare this pair truncation method and the one used in the `SNLIBERTDataset` class. What are their pros and cons?

Discussions[249].

# 17 Reinforcement Learning

**Pratik Chaudhari** (*University of Pennsylvania and Amazon*), **Rasool Fakoor** (*Amazon*), and **Kavosh Asadi** (*Amazon*)

Reinforcement Learning (RL) is a suite of techniques that allows us to build machine learning systems that take decisions sequentially. For example, a package containing new clothes that you purchased from an online retailer arrives at your doorstep after a sequence of decisions, e.g., the retailer finding the clothes in the warehouse closest to your house, putting the clothes in a box, transporting the box via land or by air, and delivering it to your house within the city. There are many variables that affect the delivery of the package along the way, e.g., whether or not the clothes were available in the warehouse, how long it took to transport the box, whether it arrived in your city before the daily delivery truck left, etc. The key idea is that at each stage these variables that we do not often control affect the entire sequence of events in the future, e.g., if there were delays in packing the box in the warehouse the retailer may need to send the package via air instead of ground to ensure a timely delivery. Reinforcement Learning methods allow us to take the appropriate action at each stage of a sequential decision making problem in order to maximize some utility eventually, e.g., the timely delivery of the package to you.

Such sequential decision making problems are seen in numerous other places, e.g., while playing Go[250] your current move determines the next moves and the opponent's moves are the variables that you cannot control… a sequence of moves eventually determines whether or not you win; the movies that Netflix recommends to you now determine what you watch, whether you like the movie or not is unknown to Netflix, eventually a sequence of movie recommendations determines how satisfied you are with Netflix. Reinforcement learning is being used today to develop effective solutions to these problems (Mnih *et al.*, 2013, Silver *et al.*, 2016). The key distinction between reinforcement learning and standard deep learning is that in standard deep learning the prediction of a trained model on one test datum does not affect the predictions on a future test datum; in reinforcement learning decisions at future instants (in RL, decisions are also called actions) are affected by what decisions were made in the past.

In this chapter, we will develop the fundamentals of reinforcement learning and obtain hands-on experience in implementing some popular reinforcement learning methods. We will first develop a concept called a Markov Decision Process (MDP) which allows us to think of such sequential decision making problems. An algorithm called Value Iteration will be our first insight into solving reinforcement learning problems under the assumption that we know how the uncontrolled variables in an MDP (in RL, these controlled variables are called the environment) typically behave. Using the more general version of Value Iteration, an algorithm called Q-Learning, we will be able to take appropriate actions even when we do not necessarily have full knowledge of the environment. We will then study how to use deep networks for reinforcement learning problems by imitating the actions of an expert. And finally, we will develop a reinforcement learning method that uses a deep network to take actions in unknown environments. These techniques form the basis of more advanced RL algorithms that are used today in a variety of real-world applications, some of which we will point to in the chapter.

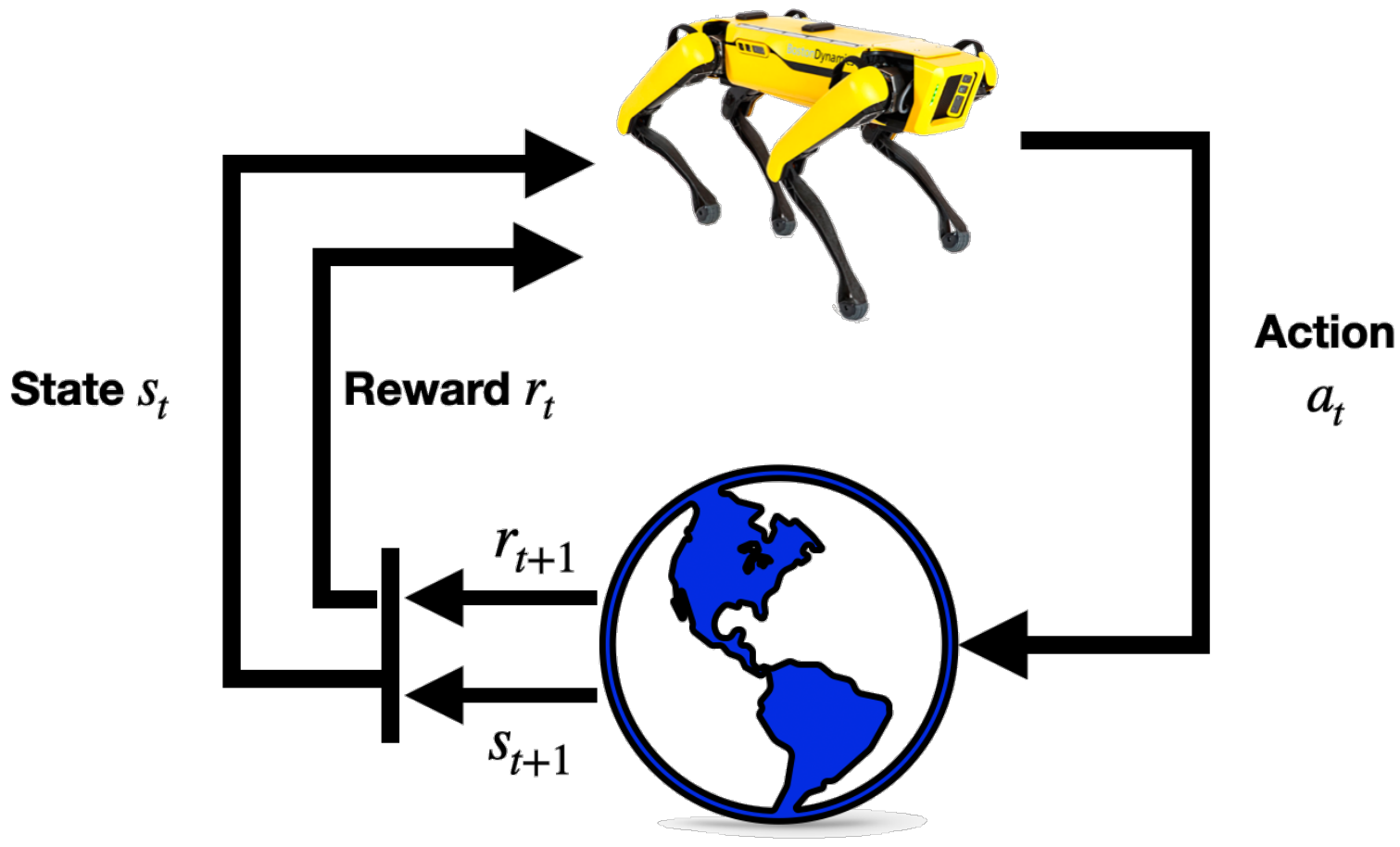

Fig. 17.1 Reinforcement Learning Structure

## 17.1 Markov Decision Process (MDP)

In this section, we will discuss how to formulate reinforcement learning problems using Markov decision processes (MDPs) and describe various components of MDPs in detail.

### 17.1.1 Definition of an MDP

A Markov decision process (MDP) (Bellman, 1957) is a model for how the state of a system evolves as different actions are applied to the system. A few different quantities come together to form an MDP.

- Let $\mathcal{S}$ be the set of states in the MDP. As a concrete example see Fig. 17.1.1, for a robot that is navigating a gridworld. In this case, $\mathcal{S}$ corresponds to the set of locations that the robot can be at any given timestep.

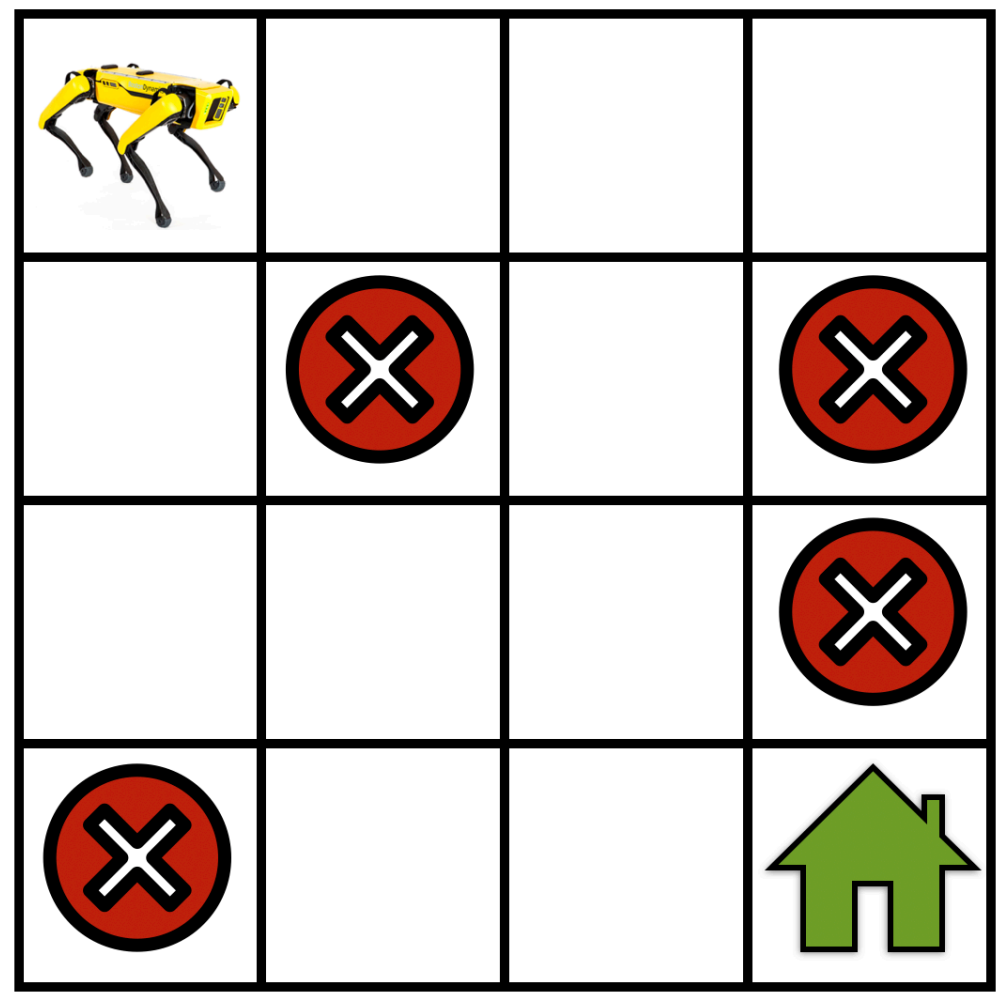

Fig. 17.1.1 A simple gridworld navigation task where the robot not only has to find its way to the goal location (shown as a green house) but also has to avoid trap locations (shown as red cross signs).

- Let $\mathcal{A}$ be the set of actions that the robot can take at each state, e.g., "go forward", "turn right", "turn left", "stay at the same location", etc. Actions can change the current state of the robot to some other state within the set $\mathcal{S}$.

- It may happen that we do not know how the robot moves *exactly* but only know it up to some approximation. We model this situation in reinforcement learning as follows: if the robot takes an action "go forward", there might be a small probability that it stays at the current state, another small probability that it "turns left", etc. Mathematically, this amounts to defining a "transition function" $T : \mathcal{S} \times \mathcal{A} \times \mathcal{S} \to [0, 1]$ such that $T(s, a, s') = P(s' \mid s, a)$ using the conditional probability of reaching a state $s'$ given that the robot was at state $s$ and took an action $a$. The transition function is a probability distribution and we therefore have $\sum_{s' \in \mathcal{S}} T(s, a, s') = 1$ for all $s \in \mathcal{S}$ and $a \in \mathcal{A}$, i.e., the robot has to go to some state if it takes an action.

- We now construct a notion of which actions are useful and which ones are not using the concept of a "reward" $r : \mathcal{S} \times \mathcal{A} \to \mathbb{R}$. We say that the robot gets a reward $r(s, a)$ if the robot takes an action $a$ at state $s$. If the reward $r(s, a)$ is large, this indicates that taking the action $a$ at state $s$ is more useful to achieving the goal of the robot, i.e., going to the green house. If the reward $r(s, a)$ is small, then action $a$ is less useful to achieving this goal. It is important to note that the reward is designed by the user (the person who creates the reinforcement learning algorithm) with the goal in mind.

### 17.1.2 Return and Discount Factor

The different components above together form a Markov decision process (MDP)

$$\text{MDP} : (\mathcal{S}, \mathcal{A}, T, r). \tag{17.1.1}$$

Let's now consider the situation when the robot starts at a particular state $s_0 \in \mathcal{S}$ and continues taking actions to result in a trajectory

$$\tau = (s_0, a_0, r_0, s_1, a_1, r_1, s_2, a_2, r_2, \ldots). \tag{17.1.2}$$

At each time step $t$ the robot is at a state $s_t$ and takes an action $a_t$ which results in a reward $r_t = r(s_t, a_t)$. The *return* of a trajectory is the total reward obtained by the robot along such a trajectory

$$R(\tau) = r_0 + r_1 + r_2 + \cdots . \tag{17.1.3}$$

The goal in reinforcement learning is to find a trajectory that has the largest *return*.

Think of the situation when the robot continues to travel in the gridworld without ever reaching the goal location. The sequence of states and actions in a trajectory can be infinitely long in this case and the *return* of any such infinitely long trajectory will be infinite. In order to keep the reinforcement learning formulation meaningful even for such trajectories, we introduce the notion of a discount factor $\gamma < 1$. We write the discounted *return* as

$$R(\tau) = r_0 + \gamma r_1 + \gamma^2 r_2 + \cdots = \sum_{t=0}^{\infty} \gamma^t r_t. \tag{17.1.4}$$

Note that if $\gamma$ is very small, the rewards earned by the robot in the far future, say $t = 1000$, are heavily discounted by the factor $\gamma^{1000}$. This encourages the robot to select short trajectories that achieve its goal, namely that of going to the green house in the gridwold example (see Fig. 17.1.1). For large values of the discount factor, say $\gamma = 0.99$, the robot is encouraged to *explore* and then find the best trajectory to go to the goal location.

### 17.1.3 Discussion of the Markov Assumption

Let us think of a new robot where the state $s_t$ is the location as above but the action $a_t$ is the acceleration that the robot applies to its wheels instead of an abstract command like "go forward". If this robot has some non-zero velocity at state $s_t$, then the next location $s_{t+1}$ is a function of the past location $s_t$, the acceleration $a_t$, also the velocity of the robot at time $t$ which is proportional to $s_t - s_{t-1}$. This indicates that we should have

$$s_{t+1} = \text{some function}(s_t, a_t, s_{t-1}); \tag{17.1.5}$$

the "some function" in our case would be Newton's law of motion. This is quite different from our transition function that simply depends upon $s_t$ and $a_t$.

Markov systems are all systems where the next state $s_{t+1}$ is only a function of the current state $s_t$ and the action $a_t$ taken at the current state. In Markov systems, the next state does not depend on which actions were taken in the past or the states that the robot was at in the past. For example, the new robot that has acceleration as the action above is not Markovian because the next location $s_{t+1}$ depends upon the previous state $s_{t-1}$ through the velocity. It may seem that Markovian nature of a system is a restrictive assumption, but it is not so. Markov Decision Processes are still capable of modeling a very large class of real systems. For example, for our new robot, if we chose our state $s_t$ to the tuple (location, velocity) then the system is Markovian because its next state $(\text{location}_{t+1}, \text{velocity}_{t+1})$ depends only upon the current state $(\text{location}_t, \text{velocity}_t)$ and the action at the current state $a_t$.

### 17.1.4 Summary

The reinforcement learning problem is typically modeled using Markov Decision Processes. A Markov decision process (MDP) is defined by a tuple of four entities $(\mathcal{S}, \mathcal{A}, T, r)$ where $\mathcal{S}$ is the state space, $\mathcal{A}$ is the action space, $T$ is the transition function that encodes the transition probabilities of the MDP and $r$ is the immediate reward obtained by taking action at a particular state.

### 17.1.5 Exercises

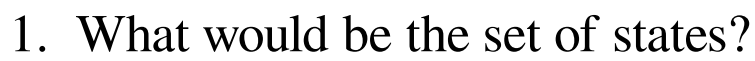

1. Suppose that we want to design an MDP to model MountainCar[251] problem.
    1. What would be the set of states?
    2. What would be the set of actions?
    3. What would be the possible reward functions?

2. How would you design an MDP for an Atari game like Pong game[252]?

Discussions[253].

## 17.2 Value Iteration

In this section we will discuss how to pick the best action for the robot at each state to maximize the *return* of the trajectory. We will describe an algorithm called Value Iteration and implement it for a simulated robot that travels over a frozen lake.

### 17.2.1 Stochastic Policy

A stochastic policy denoted as $\pi(a \mid s)$ (policy for short) is a conditional distribution over the actions $a \in \mathcal{A}$ given the state $s \in \mathcal{S}$, $\pi(a \mid s) \equiv P(a \mid s)$. As an example, if the robot has four actions $\mathcal{A} = \{\text{go left, go down, go right, go up}\}$. The policy at a state $s \in \mathcal{S}$ for such a set of actions $\mathcal{A}$ is a categorical distribution where the probabilities of the four actions could be $[0.4, 0.2, 0.1, 0.3]$; at some other state $s' \in \mathcal{S}$ the probabilities $\pi(a \mid s')$ of the same four actions could be $[0.1, 0.1, 0.2, 0.6]$. Note that we should have $\sum_a \pi(a \mid s) = 1$ for any state $s$. A deterministic policy is a special case of a stochastic policy in that the distribution $\pi(a \mid s)$ only gives non-zero probability to one particular action, e.g., $[1, 0, 0, 0]$ for our example with four actions.

To make the notation less cumbersome, we will often write $\pi(s)$ as the conditional distribution instead of $\pi(a \mid s)$.

### 17.2.2 Value Function

Imagine now that the robot starts at a state $s_0$ and at each time instant, it first samples an action from the policy $a_t \sim \pi(s_t)$ and takes this action to result in the next state $s_{t+1}$. The trajectory $\tau = (s_0, a_0, r_0, s_1, a_1, r_1, \ldots)$, can be different depending upon which particular action $a_t$ is sampled at intermediate instants. We define the average *return* $R(\tau) = \sum_{t=0}^{\infty} \gamma^t r(s_t, a_t)$ of all such trajectories

$$V^{\pi}(s_0) = E_{a_t \sim \pi(s_t)} \Big[ R(\tau) \Big] = E_{a_t \sim \pi(s_t)} \Big[ \sum_{t=0}^{\infty} \gamma^t r(s_t, a_t) \Big], \tag{17.2.1}$$

where $s_{t+1} \sim P(s_{t+1} \mid s_t, a_t)$ is the next state of the robot and $r(s_t, a_t)$ is the instantaneous reward obtained by taking action $a_t$ in state $s_t$ at time $t$. This is called the "value function" for the policy $\pi$. In simple words, the value of a state $s_0$ for a policy $\pi$, denoted by $V^{\pi}(s_0)$, is the expected $\gamma$-discounted *return* obtained by the robot if it begins at state $s_0$ and takes actions from the policy $\pi$ at each time instant.

We next break down the trajectory into two stages (i) the first stage which corresponds to $s_0 \to s_1$ upon taking the action $a_0$, and (ii) a second stage which is the trajectory $\tau' = (s_1, a_1, r_1, \ldots)$ thereafter. The key idea behind all algorithms in reinforcement learning is that the value of state $s_0$ can be written as the average reward obtained in the first stage and the value function averaged over all possible next states $s_1$. This is quite intuitive and arises from our Markov assumption: the average return from the current state is the sum of the average return from the next state and the average reward of going to the next state. Mathematically, we write the two stages as

$$V^{\pi}(s_0) = r(s_0, a_0) + \gamma\ E_{a_0 \sim \pi(s_0)} \Big[ E_{s_1 \sim P(s_1 \mid s_0, a_0)} \Big[ V^{\pi}(s_1) \Big] \Big]. \tag{17.2.2}$$

This decomposition is very powerful: it is the foundation of the principle of dynamic programming upon which all reinforcement learning algorithms are based. Notice that the second stage gets two expectations, one over the choices of the action $a_0$ taken in the first stage using the stochastic policy and another over the possible states $s_1$ obtained from the chosen action. We can write (17.2.2) using the transition probabilities in the Markov decision process (MDP) as

$$V^{\pi}(s) = \sum_{a \in \mathcal{A}} \pi(a \mid s) \Big[ r(s, a) + \gamma \sum_{s' \in \mathcal{S}} P(s' \mid s, a) V^{\pi}(s') \Big];\ \text{for all } s \in \mathcal{S}. \tag{17.2.3}$$

An important thing to notice here is that the above identity holds for all states $s \in \mathcal{S}$ because we can think of any trajectory that begins at that state and break down the trajectory into two stages.

### 17.2.3 Action-Value Function

In implementations, it is often useful to maintain a quantity called the "action value" function which is a closely related quantity to the value function. This is defined to be the average *return* of a trajectory that begins at $s_0$ but when the action of the first stage is fixed to be $a_0$

$$Q^\pi(s_0, a_0) = r(s_0, a_0) + E_{a_t \sim \pi(s_t)} \Big[ \sum_{t=1}^\infty \gamma^t r(s_t, a_t) \Big], \tag{17.2.4}$$

note that the summation inside the expectation is from $t = 1, \ldots, \infty$ because the reward of the first stage is fixed in this case. We can again break down the trajectory into two parts and write

$$Q^\pi(s, a) = r(s, a) + \gamma \sum_{s' \in \mathcal{S}} P(s' \mid s, a) \sum_{a' \in \mathcal{A}} \pi(a' \mid s') \; Q^\pi(s', a'); \;\; \text{for all } s \in \mathcal{S}, a \in \mathcal{A}. \tag{17.2.5}$$

This version is the analog of (17.2.3) for the action value function.

### 17.2.4 Optimal Stochastic Policy

Both the value function and the action-value function depend upon the policy that the robot chooses. We will next think of the "optimal policy" that achieves the maximal average *return*

$$\pi^* = \underset{\pi}{\mathrm{argmax}} V^\pi(s_0). \tag{17.2.6}$$

Of all possible stochastic policies that the robot could have taken, the optimal policy $\pi^*$ achieves the largest average discounted *return* for trajectories starting from state $s_0$. Let us denote the value function and the action-value function of the optimal policy as $V^* \equiv V^{\pi^*}$ and $Q^* \equiv Q^{\pi^*}$.

Let us observe that for a deterministic policy where there is only one action that is possible under the policy at any given state. This gives us

$$\pi^*(s) = \underset{a \in \mathcal{A}}{\mathrm{argmax}} \Big[ r(s, a) + \gamma \sum_{s' \in \mathcal{S}} P(s' \mid s, a) \; V^*(s') \Big]. \tag{17.2.7}$$

A good mnemonic to remember this is that the optimal action at state $s$ (for a deterministic policy) is the one that maximizes the sum of reward $r(s, a)$ from the first stage and the average *return* of the trajectories starting from the next sate $s'$, averaged over all possible next states $s'$ from the second stage.

### 17.2.5 Principle of Dynamic Programming

Our developement in the previous section in (17.2.2) or (17.2.5) can be turned into an algorithm to compute the optimal value function $V^*$ or the action-value function $Q^*$, respectively. Observe that

$$V^*(s) = \sum_{a\in\mathcal{A}} \pi^*(a \mid s)\Big[r(s,a) + \gamma \sum_{s'\in\mathcal{S}} P(s' \mid s,a)V^*(s')\Big]; \text{ for all } s \in \mathcal{S}. \tag{17.2.8}$$

For a deterministic optimal policy $\pi^*$, since there is only one action that can be taken at state $s$, we can also write

$$V^*(s) = \mathrm{argmax}_{a\in\mathcal{A}}\Big\{r(s,a) + \gamma \sum_{s'\in\mathcal{S}} P(s' \mid s,a)V^*(s')\Big\} \tag{17.2.9}$$

for all states $s \in \mathcal{S}$. This identity is called the "principle of dynamic programming" (Bellman, 1952, Bellman, 1957). It was formulated by Richard Bellman in 1950s and we can remember it as "the remainder of an optimal trajectory is also optimal".

### 17.2.6 Value Iteration

We can turn the principle of dynamic programming into an algorithm for finding the optimal value function called value iteration. The key idea behind value iteration is to think of this identity as a set of constraints that tie together $V^*(s)$ at different states $s \in \mathcal{S}$. We initialize the value function to some arbitrary values $V_0(s)$ for all states $s \in \mathcal{S}$. At the $k^{\text{th}}$ iteration, the Value Iteration algorithm updates the value function as

$$V_{k+1}(s) = \max_{a\in\mathcal{A}}\Big\{r(s,a) + \gamma \sum_{s'\in\mathcal{S}} P(s' \mid s,a)V_k(s')\Big\}; \text{ for all } s \in \mathcal{S}. \tag{17.2.10}$$

It turns out that as $k \to \infty$ the value function estimated by the Value Iteration algorithm converges to the optimal value function irrespective of the initialization $V_0$,

$$V^*(s) = \lim_{k\to\infty} V_k(s); \text{ for all states } s \in \mathcal{S}. \tag{17.2.11}$$

The same Value Iteration algorithm can be equivalently written using the action-value function as

$$Q_{k+1}(s,a) = r(s,a) + \gamma \max_{a'\in\mathcal{A}} \sum_{s'\in\mathcal{S}} P(s' \mid s,a)Q_k(s',a'); \text{ for all } s \in \mathcal{S}, a \in \mathcal{A}. \tag{17.2.12}$$

In this case we initialize $Q_0(s,a)$ to some arbitrary values for all $s \in \mathcal{S}$ and $a \in \mathcal{A}$. Again we have $Q^*(s,a) = \lim_{k\to\infty} Q_k(s,a)$ for all $s \in \mathcal{S}$ and $a \in \mathcal{A}$.

### 17.2.7 Policy Evaluation

Value Iteration enables us to compute the optimal value function, i.e., $V^{\pi^*}$ of the optimal deterministic policy $\pi^*$. We can also use similar iterative updates to compute the value function associated with any other, potentially stochastic, policy $\pi$. We again initialize $V_0^\pi(s)$ to some arbitrary values for all states $s \in \mathcal{S}$ and at the $k^{\text{th}}$ iteration, perform the updates

$$V_{k+1}^\pi(s) = \sum_{a \in \mathcal{A}} \pi(a \mid s) \Big[ r(s,a) + \gamma \sum_{s' \in \mathcal{S}} P(s' \mid s, a) V_k^\pi(s') \Big]; \text{ for all } s \in \mathcal{S}. \tag{17.2.13}$$

This algorithm is known as policy evaluation and is useful to compute the value function given the policy. Again, it turns out that as $k \to \infty$ these updates converge to the correct value function irrespective of the initialization $V_0$,

$$V^\pi(s) = \lim_{k \to \infty} V_k^\pi(s); \text{ for all states } s \in \mathcal{S}. \tag{17.2.14}$$

The algorithm for computing the action-value function $Q^\pi(s,a)$ of a policy $\pi$ is analogous.

## 17.2.8 Implementation of Value Iteration

We next show how to implement Value Iteration for a navigation problem called FrozenLake from Open AI Gym[254]. We first need to setup the enviroment as shown in the following code.

```
%matplotlib inline
import random
import numpy as np
from d2l import torch as d2l

seed = 0  # Random number generator seed
gamma = 0.95  # Discount factor
num_iters = 10  # Number of iterations
random.seed(seed)  # Set the random seed to ensure results can be reproduced
np.random.seed(seed)

# Now set up the environment
env_info = d2l.make_env('FrozenLake-v1', seed=seed)
```

In the FrozenLake environment, the robot moves on a $4 \times 4$ grid (these are the states) with actions that are “up” (↑), “down” (→), “left” (←), and “right” (→). The environment contains a number of holes (H) cells and frozen (F) cells as well as a goal cell (G), all of which are unknown to the robot. To keep the problem simple, we assume the robot has reliable actions, i.e. $P(s' \mid s, a) = 1$ for all $s \in \mathcal{S}, a \in \mathcal{A}$. If the robot reaches the goal, the trial ends and the robot receives a reward of $1$ irrespective of the action; the reward at any other state is $0$ for all actions. The objective of the robot is to learn a policy that reaches the goal location (G) from a given start location (S) (this is $s_0$) to maximize the *return*.

The following function implements Value Iteration, where `env_info` contains MDP and environment related information and `gamma` is the discount factor:

```
def value_iteration(env_info, gamma, num_iters):
    env_desc = env_info['desc']  # 2D array shows what each item means
    prob_idx = env_info['trans_prob_idx']
```

```
    nextstate_idx = env_info['nextstate_idx']
    reward_idx = env_info['reward_idx']
    num_states = env_info['num_states']
    num_actions = env_info['num_actions']
    mdp = env_info['mdp']

    V  = np.zeros((num_iters + 1, num_states))
    Q  = np.zeros((num_iters + 1, num_states, num_actions))
    pi = np.zeros((num_iters + 1, num_states))

    for k in range(1, num_iters + 1):
        for s in range(num_states):
            for a in range(num_actions):
                # Calculate \sum_{s'} p(s'\mid s,a) [r + \gamma v_k(s')]
                for pxrds in mdp[(s,a)]:
                    # mdp(s,a): [(p1,next1,r1,d1),(p2,next2,r2,d2),..]
                    pr = pxrds[prob_idx]  # p(s'\mid s,a)
                    nextstate = pxrds[nextstate_idx]  # Next state
                    reward = pxrds[reward_idx]  # Reward
                    Q[k,s,a] += pr * (reward + gamma * V[k - 1, nextstate])
            # Record max value and max action
            V[k,s] = np.max(Q[k,s,:])
            pi[k,s] = np.argmax(Q[k,s,:])
    d2l.show_value_function_progress(env_desc, V[:-1], pi[:-1])

value_iteration(env_info=env_info, gamma=gamma, num_iters=num_iters)
```

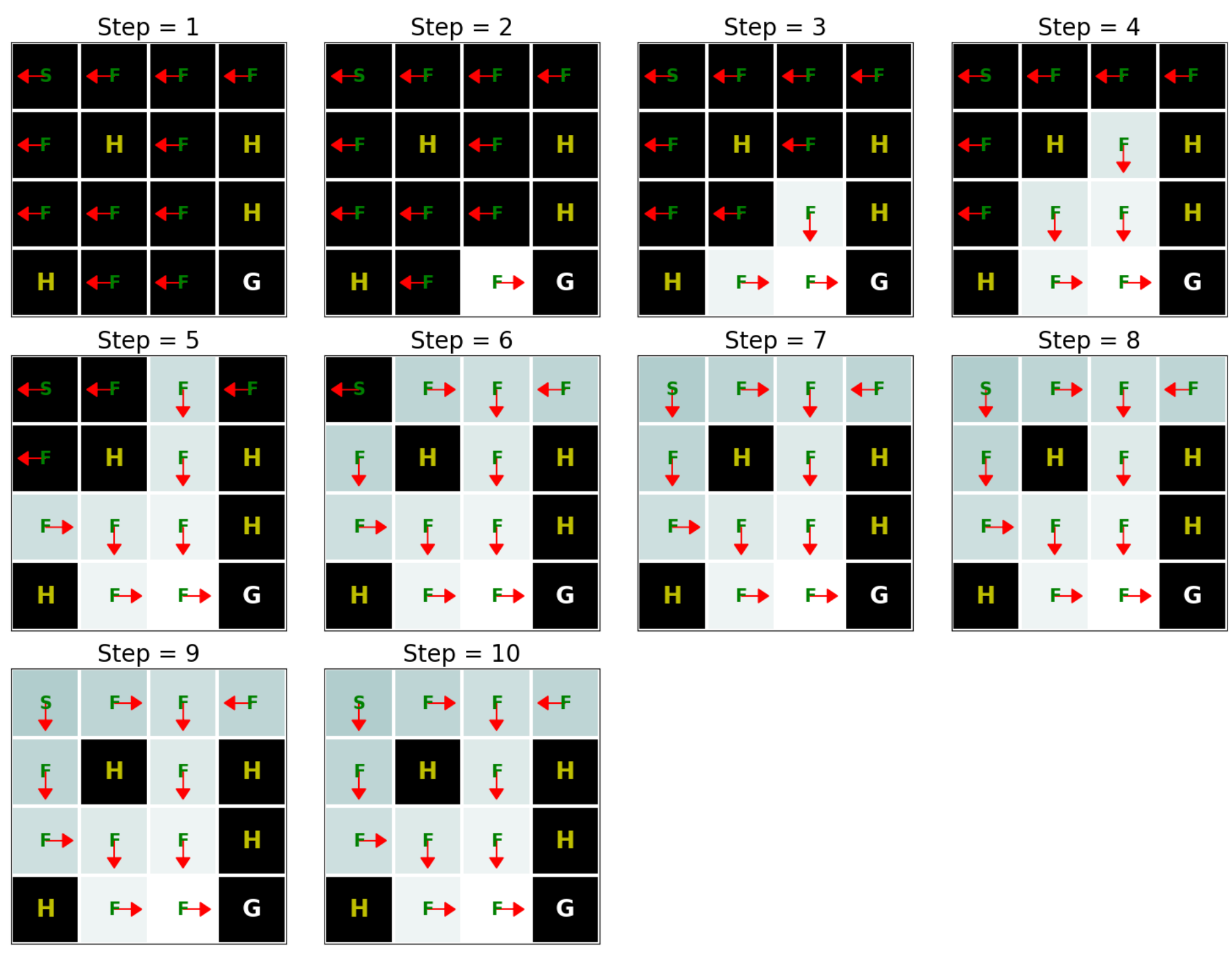

The above pictures show the policy (the arrow indicates the action) and value function (the change in color shows how the value function changes over time from the initial value shown by dark color to the optimal value shown by light colors.). As we see, Value Iteration finds the optimal value function after 10 iterations and the goal state (G) can be reached starting from any state as long as it is not an H cell. Another interesting aspect of the implementation is that in addition to finding the optimal value function, we also automatically found the optimal policy $\pi^*$ corresponding to this value function.

### 17.2.9 Summary

The main idea behind the Value Iteration algorithm is to use the principle of dynamic programming to find the optimal average return obtained from a given state. Note that implementing the Value Iteration algorithm requires that we know the Markov decision process (MDP), e.g., the transition and reward functions, completely.

### 17.2.10 Exercises

1. Try increasing the grid size to $8 \times 8$. Compared with $4 \times 4$ grid, how many iterations does it take to find the optimal value function?
2. What is the computational complexity of the Value Iteration algorithm?
3. Run the Value Iteration algorithm again with $\gamma$ (i.e. "gamma" in the above code) when it equals to $0$, $0.5$, and $1$ and analyze its results.
4. How does the value of $\gamma$ affect the number of iterations taken by Value Iteration to converge? What happens when $\gamma = 1$?

Discussions[255].

## 17.3 Q-Learning

In the previous section, we discussed the Value Iteration algorithm which requires accessing the complete Markov decision process (MDP), e.g., the transition and reward functions. In this section, we will look at Q-Learning (Watkins and Dayan, 1992) which is an algorithm to learn the value function without necessarily knowing the MDP. This algorithm embodies the central idea behind reinforcement learning: it will enable the robot to obtain its own data.

### 17.3.1 The Q-Learning Algorithm

Recall that value iteration for the action-value function in *Value Iteration* (page 813) corresponds to the update

$$Q_{k+1}(s,a) = r(s,a) + \gamma \sum_{s' \in \mathcal{S}} P(s' \mid s,a) \max_{a' \in \mathcal{A}} Q_k(s',a'); \text{ for all } s \in \mathcal{S} \text{ and } a \in \mathcal{A}. \tag{17.3.1}$$

As we discussed, implementing this algorithm requires knowing the MDP, specifically the transition function $P(s' \mid s,a)$. The key idea behind Q-Learning is to replace the summation over all $s' \in \mathcal{S}$ in the above expression by a summation over the states visited by the robot. This allows us to subvert the need to know the transition function.

### 17.3.2 An Optimization Problem Underlying Q-Learning

Let us imagine that the robot uses a policy $\pi_e(a \mid s)$ to take actions. Just like the previous chapter, it collects a dataset of $n$ trajectories of $T$ timesteps each $\{(s_t^i, a_t^i)_{t=0,\ldots,T-1}\}_{i=1,\ldots,n}$. Recall that value iteration is really a set of constraints that ties together the action-value $Q^*(s,a)$ of different states and actions to each other. We can implement an approximate version of value iteration using the data that the robot has collected using $\pi_e$ as

$$\hat{Q} = \min_Q \underbrace{\frac{1}{nT} \sum_{i=1}^{n} \sum_{t=0}^{T-1} (Q(s_t^i, a_t^i) - r(s_t^i, a_t^i) - \gamma \max_{a'} Q(s_{t+1}^i, a'))^2}_{\stackrel{\text{def}}{=} \ell(Q)}. \tag{17.3.2}$$

Let us first observe the similarities and differences between this expression and value iteration above. If the robot's policy $\pi_e$ were equal to the optimal policy $\pi^*$, and if it collected an infinite amount of data, then this optimization problem would be identical to the optimization problem underlying value iteration. But while value iteration requires us to know $P(s' \mid s,a)$, the optimization objective does not have this term. We have not cheated: as the robot uses the policy $\pi_e$ to take an action $a_t^i$ at state $s_t^i$, the next state $s_{t+1}^i$ is a sample drawn from the transition function. So the optimization objective also has access to the transition function, but implicitly in terms of the data collected by the robot.

The variables of our optimization problem are $Q(s,a)$ for all $s \in \mathcal{S}$ and $a \in \mathcal{A}$. We can minimize the objective using gradient descent. For every pair $(s_t^i, a_t^i)$ in our dataset, we can write

$$\begin{aligned} Q(s_t^i, a_t^i) &\leftarrow Q(s_t^i, a_t^i) - \alpha \nabla_{Q(s_t^i, a_t^i)} \ell(Q) \\ &= (1-\alpha) Q(s_t^i, a_t^i) - \alpha \Big( r(s_t^i, a_t^i) + \gamma \max_{a'} Q(s_{t+1}^i, a') \Big), \end{aligned} \tag{17.3.3}$$

where $\alpha$ is the learning rate. Typically in real problems, when the robot reaches the goal location, the trajectories end. The value of such a terminal state is zero because the robot does not take any further actions beyond this state. We should modify our update to handle

such states as

$$Q(s_t^i, a_t^i) = (1-\alpha)Q(s_t^i, a_t^i) - \alpha\Big(r(s_t^i, a_t^i) + \gamma(1 - \mathbb{1}_{s_{t+1}^i \text{ is terminal}}) \max_{a'} Q(s_{t+1}^i, a')\Big). \tag{17.3.4}$$

where $\mathbb{1}_{s_{t+1}^i \text{ is terminal}}$ is an indicator variable that is one if $s_{t+1}^i$ is a terminal state and zero otherwise. The value of state-action tuples $(s, a)$ that are not a part of the dataset is set to $-\infty$. This algorithm is known as Q-Learning.

Given the solution of these updates $\hat{Q}$, which is an approximation of the optimal value function $Q^*$, we can obtain the optimal deterministic policy corresponding to this value function easily using

$$\hat{\pi}(s) = \mathrm{argmax}_a \hat{Q}(s, a). \tag{17.3.5}$$

There can be situations when there are multiple deterministic policies that correspond to the same optimal value function; such ties can be broken arbitrarily because they have the same value function.

### 17.3.3 Exploration in Q-Learning

The policy used by the robot to collect data $\pi_e$ is critical to ensure that Q-Learning works well. Afterall, we have replaced the expectation over $s'$ using the transition function $P(s' \mid s, a)$ using the data collected by the robot. If the policy $\pi_e$ does not reach diverse parts of the state-action space, then it is easy to imagine our estimate $\hat{Q}$ will be a poor approximation of the optimal $Q^*$. It is also important to note that in such a situation, the estimate of $Q^*$ at *all states* $s \in \mathcal{S}$ will be bad, not just the ones visited by $\pi_e$. This is because the Q-Learning objective (or value iteration) is a constraint that ties together the value of all state-action pairs. It is therefore critical to pick the correct policy $\pi_e$ to collect data.

We can mitigate this concern by picking a completely random policy $\pi_e$ that samples actions uniformly randomly from $\mathcal{A}$. Such a policy would visit all states, but it will take a large number of trajectories before it does so.

We thus arrive at the second key idea in Q-Learning, namely exploration. Typical implementations of Q-Learning tie together the current estimate of $Q$ and the policy $\pi_e$ to set

$$\pi_e(a \mid s) = \begin{cases} \mathrm{argmax}_{a'} \hat{Q}(s, a') & \text{with prob. } 1-\epsilon \\ \text{uniform}(\mathcal{A}) & \text{with prob. } \epsilon, \end{cases} \tag{17.3.6}$$

where $\epsilon$ is called the "exploration parameter" and is chosen by the user. The policy $\pi_e$ is called an exploration policy. This particular $\pi_e$ is called an $\epsilon$-greedy exploration policy because it chooses the optimal action (under the current estimate $\hat{Q}$) with probability $1-\epsilon$ but explores randomly with the remainder probability $\epsilon$. We can also use the so-called softmax exploration policy

$$\pi_e(a \mid s) = \frac{e^{\hat{Q}(s,a)/T}}{\sum_{a'} e^{\hat{Q}(s,a')/T}}; \tag{17.3.7}$$

where the hyper-parameter $T$ is called temperature. A large value of $\epsilon$ in $\epsilon$-greedy policy functions similarly to a large value of temperature $T$ for the softmax policy.

It is important to note that when we pick an exploration that depends upon the current estimate of the action-value function $\hat{Q}$, we need to resolve the optimization problem periodically. Typical implementations of Q-Learning make one mini-batch update using a few state-action pairs in the collected dataset (typically the ones collected from the previous timestep of the robot) after taking every action using $\pi_e$.

### 17.3.4 The "Self-correcting" Property of Q-Learning

The dataset collected by the robot during Q-Learning grows with time. Both the exploration policy $\pi_e$ and the estimate $\hat{Q}$ evolve as the robot collects more data. This gives us a key insight into why Q-Learning works well. Consider a state $s$: if a particular action $a$ has a large value under the current estimate $\hat{Q}(s,a)$, then both the $\epsilon$-greedy and the softmax exploration policies have a larger probability of picking this action. If this action actually is *not* the ideal action, then the future states that arise from this action will have poor rewards. The next update of the Q-Learning objective will therefore reduce the value $\hat{Q}(s,a)$, which will reduce the probability of picking this action the next time the robot visits state $s$. Bad actions, e.g., ones whose value is overestimated in $\hat{Q}(s,a)$, are explored by the robot but their value is correct in the next update of the Q-Learning objective. Good actions, e.g., whose value $\hat{Q}(s,a)$ is large, are explored more often by the robot and thereby reinforced. This property can be used to show that Q-Learning can converge to the optimal policy even if it begins with a random policy $\pi_e$ (Watkins and Dayan, 1992).

This ability to not only collect new data but also collect the right kind of data is the central feature of reinforcement learning algorithms, and this is what distinguishes them from supervised learning. Q-Learning, using deep neural networks (which we will see in the DQN chapeter later), is responsible for the resurgence of reinforcement learning (Mnih *et al.*, 2013).

### 17.3.5 Implementation of Q-Learning

We now show how to implement Q-Learning on FrozenLake from Open AI Gym[256]. Note this is the same setup as we consider in *Value Iteration* (page 813) experiment.

```
%matplotlib inline
import random
import numpy as np
from d2l import torch as d2l

seed = 0  # Random number generator seed
gamma = 0.95  # Discount factor
num_iters = 256  # Number of iterations
alpha   = 0.9  # Learing rate
epsilon = 0.9  # Epsilon in epsilion gready algorithm
```

```
random.seed(seed)  # Set the random seed
np.random.seed(seed)

# Now set up the environment
env_info = d2l.make_env('FrozenLake-v1', seed=seed)
```

In the FrozenLake environment, the robot moves on a $4 \times 4$ grid (these are the states) with actions that are “up” ($\uparrow$), “down” ($\rightarrow$), “left” ($\leftarrow$), and “right” ($\rightarrow$). The environment contains a number of holes (H) cells and frozen (F) cells as well as a goal cell (G), all of which are unknown to the robot. To keep the problem simple, we assume the robot has reliable actions, i.e. $P(s' \mid s, a) = 1$ for all $s \in \mathcal{S}, a \in \mathcal{A}$. If the robot reaches the goal, the trial ends and the robot receives a reward of $1$ irrespective of the action; the reward at any other state is $0$ for all actions. The objective of the robot is to learn a policy that reaches the goal location (G) from a given start location (S) (this is $s_0$) to maximize the *return*.

We first implement $\epsilon$-greedy method as follows:

```
def e_greedy(env, Q, s, epsilon):
    if random.random() < epsilon:
        return env.action_space.sample()

    else:
        return np.argmax(Q[s,:])
```

We are now ready to implement Q-learning:

```
def q_learning(env_info, gamma, num_iters, alpha, epsilon):
    env_desc = env_info['desc']  # 2D array specifying what each grid item␣
↪means
    env = env_info['env']  # 2D array specifying what each grid item means
    num_states = env_info['num_states']
    num_actions = env_info['num_actions']

    Q  = np.zeros((num_states, num_actions))
    V  = np.zeros((num_iters + 1, num_states))
    pi = np.zeros((num_iters + 1, num_states))

    for k in range(1, num_iters + 1):
        # Reset environment
        state, done = env.reset(), False
        while not done:
            # Select an action for a given state and acts in env based on␣
↪selected action
            action = e_greedy(env, Q, state, epsilon)
            next_state, reward, done, _ = env.step(action)

            # Q-update:
            y = reward + gamma * np.max(Q[next_state,:])
            Q[state, action] = Q[state, action] + alpha * (y - Q[state,␣
↪action])
```

```
            # Move to the next state
            state = next_state
        # Record max value and max action for visualization purpose only
        for s in range(num_states):
            V[k,s]  = np.max(Q[s,:])
            pi[k,s] = np.argmax(Q[s,:])
    d2l.show_Q_function_progress(env_desc, V[:-1], pi[:-1])

q_learning(env_info=env_info, gamma=gamma, num_iters=num_iters, alpha=alpha,␣
↪epsilon=epsilon)
```

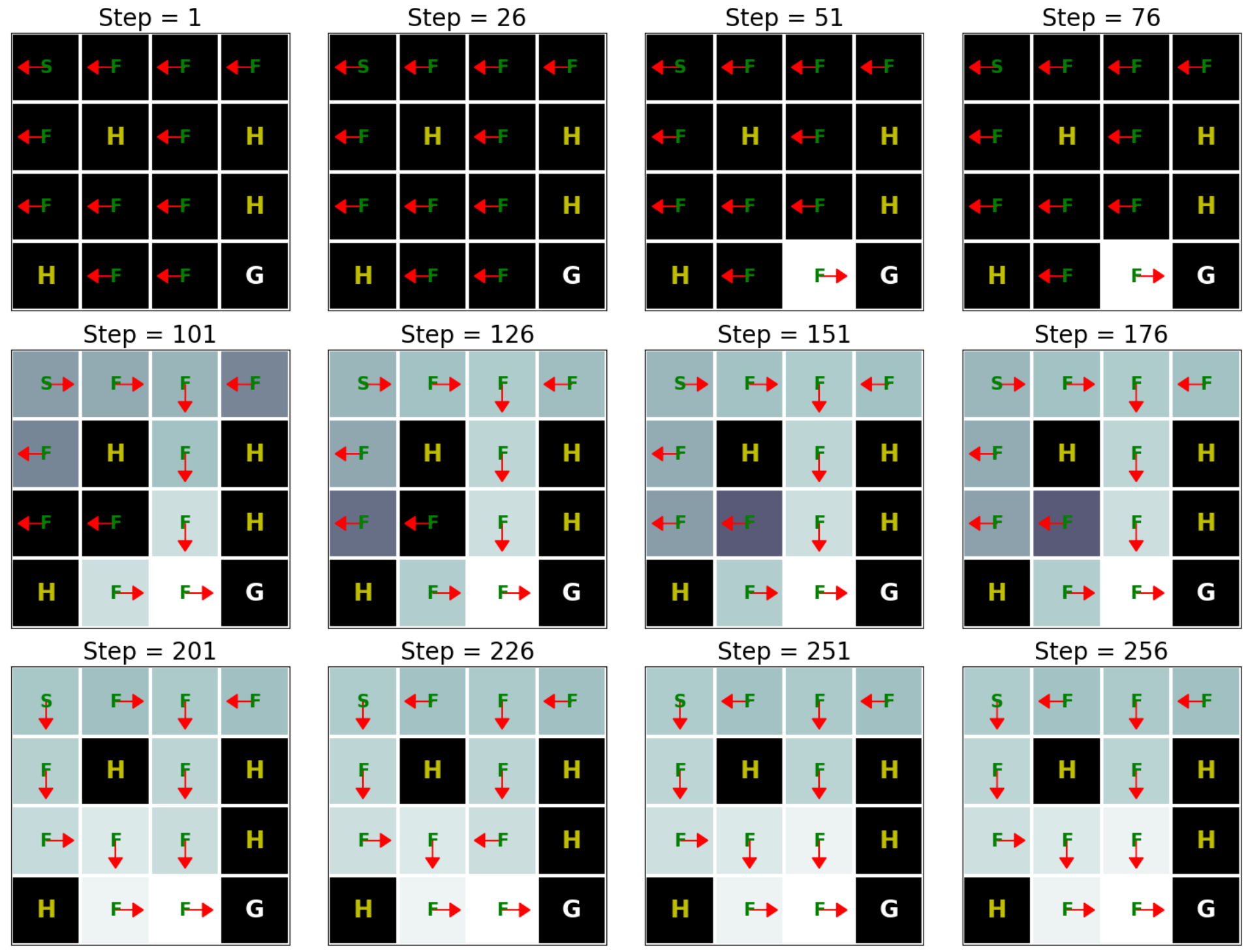

This result shows that Q-learning can find the optimal solution for this problem roughly after 250 iterations. However, when we compare this result with the Value Iteration algorithm's result (see *Implementation of Value Iteration* (page 817)), we can see that the Value Iteration algorithm needs way fewer iterations to find the optimal solution for this problem. This happens because the Value Iteration algorithm has access to the full MDP whereas Q-learning does not.

## 17.3.6 Summary

Q-learning is one of the most fundamental reinforcement-learning algorithms. It has been at the epicenter of the recent success of reinforcement learning, most notably in learning to play video games (Mnih *et al.*, 2013). Implementing Q-learning does not require that we know the Markov decision process (MDP), e.g., the transition and reward functions, completely.

### 17.3.7 Exercises

1. Try increasing the grid size to $8 \times 8$. Compared with $4 \times 4$ grid, how many iterations does it take to find the optimal value function?
2. Run the Q-learning algorithm again with $\gamma$ (i.e. "gamma" in the above code) when it equals to $0$, $0.5$, and $1$ and analyze its results.
3. Run the Q-learning algorithm again with $\epsilon$ (i.e. "epsilon" in the above code) when it equals to $0$, $0.5$, and $1$ and analyze its results.

Discussions[257].

# 18 Gaussian Processes

**Andrew Gordon Wilson** (*New York University and Amazon*)

Gaussian processes (GPs) are ubitiquous. You have already encountered many examples of GPs without realizing it. Any model that is linear in its parameters with a Gaussian distribution over the parameters is a Gaussian process. This class spans discrete models, including random walks, and autoregressive processes, as well as continuous models, including Bayesian linear regression models, polynomials, Fourier series, radial basis functions, and even neural networks with an infinite number of hidden units. There is a running joke that "everything is a special case of a Gaussian process".

Learning about Gaussian processes is important for three reasons: (1) they provide a *function space* perspective of modelling, which makes understanding a variety of model classes, including deep neural networks, much more approachable; (2) they have an extraordinary range of applications where they are state-of-the-art, including active learning, hyperparameter learning, auto-ML, and spatiotemporal regression; (3) over the last few years, algorithmic advances have made Gaussian processes increasingly scalable and relevant, harmonizing with deep learning through frameworks such as GPyTorch[258] (Gardner *et al.*, 2018). Indeed, GPs and and deep neural networks are not competing approaches, but highly complementary, and can be combined to great effect. These algorithmic advances are not just relevant to Gaussian processes, but provide a foundation in numerical methods that is broadly useful in deep learning.

In this chapter, we introduce Gaussian processes. In the introductory notebook, we start by reasoning intuitively about what Gaussian processes are and how they directly model functions. In the priors notebook, we focus on how to specify Gaussian process priors. We directly connect the tradiational weight-space approach to modelling to function space, which will help us reason about constructing and understanding machine learning models, including deep neural networks. We then introduce popular covariance functions, also known as *kernels*, which control the generalization properties of a Gaussian process. A GP with a given kernel defines a prior over functions. In the inference notebook, we will show how to use data to infer a *posterior*, in order to make predictions. This notebook contains from-scratch code for making predictions with a Gaussian process, as well as an introduction to GPyTorch. In upcoming notebooks, we will introduce the numerics behind Gaussian processes, which is useful for scaling Gaussian processes but also a powerful general foundation for deep learning, and advanced use-cases such as hyperparameter tuning in deep learning. Our examples will make use of GPyTorch, which makes Gaussian processes scale, and is closely integrated with deep learning functionality and PyTorch.

## 18.1 Introduction to Gaussian Processes

In many cases, machine learning amounts to estimating parameters from data. These parameters are often numerous and relatively uninterpretable — such as the weights of a neural network. Gaussian processes, by contrast, provide a mechanism for directly reasoning about the high-level properties of functions that could fit our data. For example, we may have a sense of whether these functions are quickly varying, periodic, involve conditional independencies, or translation invariance. Gaussian processes enable us to easily incorporate these properties into our model, by directly specifying a Gaussian distribution over the function values that could fit our data.

Let's get a feel for how Gaussian processes operate, by starting with some examples.

Suppose we observe the following dataset, of regression targets (outputs), $y$, indexed by inputs, $x$. As an example, the targets could be changes in carbon dioxide concentrations, and the inputs could be the times at which these targets have been recorded. What are some features of the data? How quickly does it seem to varying? Do we have data points collected at regular intervals, or are there missing inputs? How would you imagine filling in the missing regions, or forecasting up until $x = 25$?

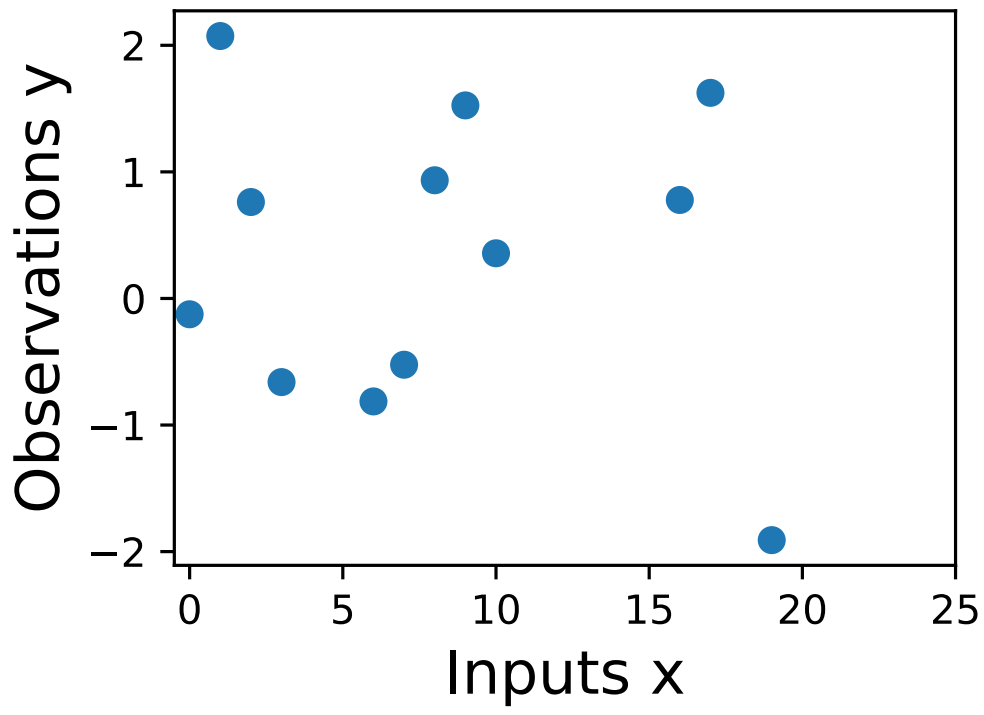

Fig. 18.1.1 Observed data.

In order to fit the data with a Gaussian process, we start by specifying a prior distribution over what types of functions we might believe to be reasonable. Here we show several sample functions from a Gaussian process. Does this prior look reasonable? Note here we are not looking for functions that fit our dataset, but instead for specifying reasonable high-level properties of the solutions, such as how quickly they vary with inputs. Note that we will see code for reproducing all of the plots in this notebook, in the next notebooks on priors and inference.

Once we condition on data, we can use this prior to infer a posterior distribution over functions that could fit the data. Here we show sample posterior functions.

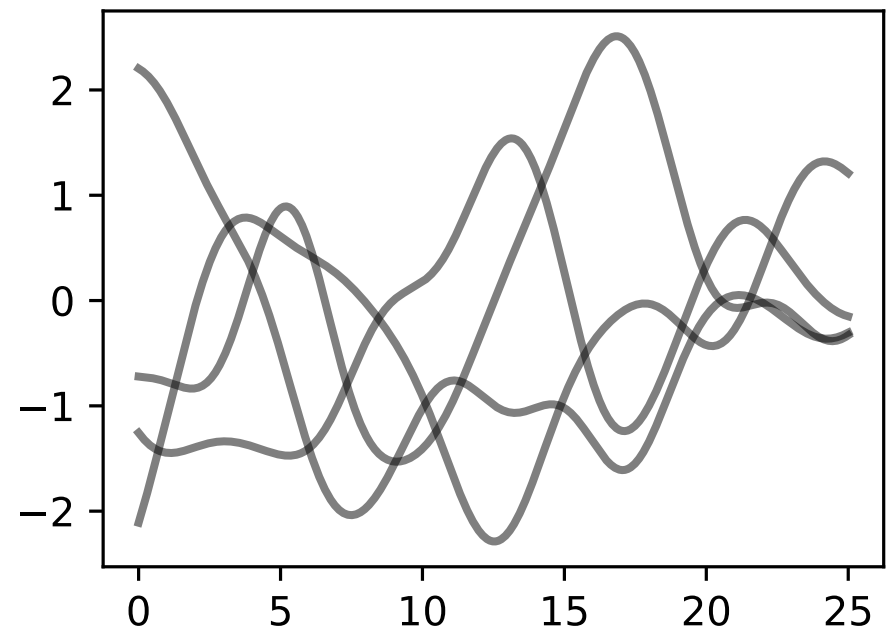

Fig. 18.1.2 Sample prior functions that we may want to represent with our model.

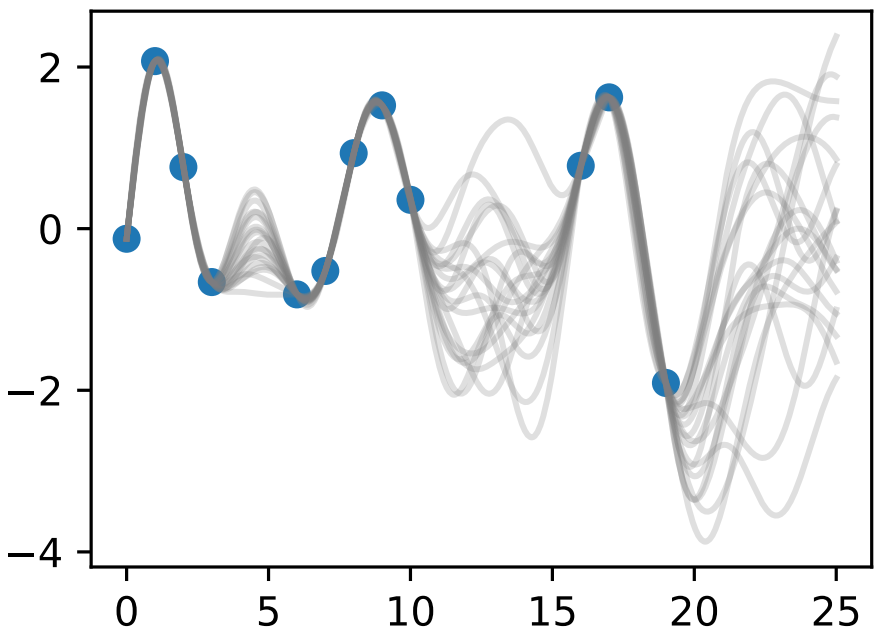

Fig. 18.1.3 Sample posterior functions, once we have observed the data.

We see that each of these functions are entirely consistent with our data, perfectly running through each observation. In order to use these posterior samples to make predictions, we can average the values of every possible sample function from the posterior, to create the curve below, in thick blue. Note that we do not actually have to take an infinite number of samples to compute this expectation; as we will see later, we can compute the expectation in closed form.

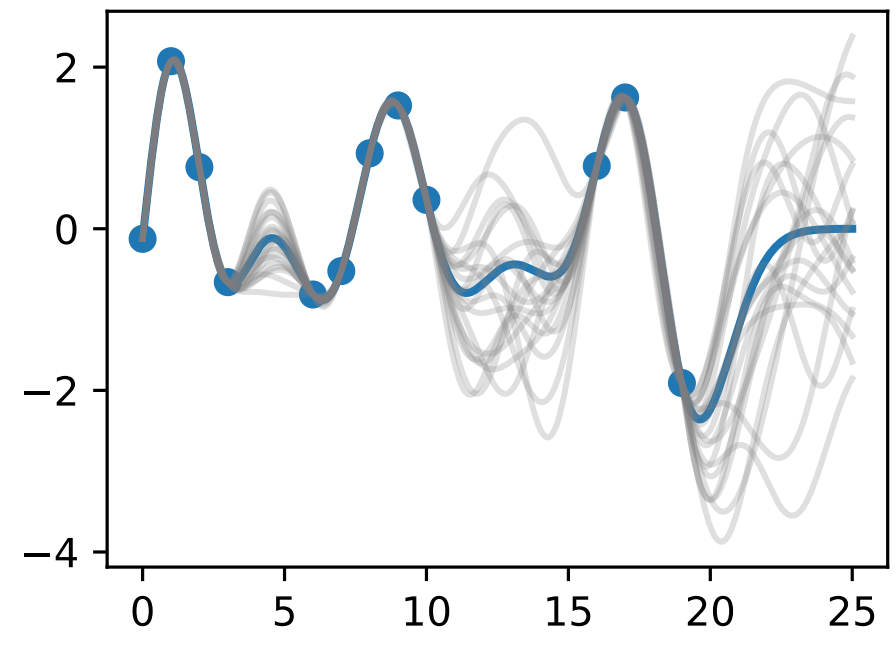

Fig. 18.1.4 Posterior samples, alongside posterior mean, which can be used for point predictions, in blue.

We may also want a representation of uncertainty, so we know how confident we should be in our predictions. Intuitively, we should have more uncertainty where there is more variability in the sample posterior functions, as this tells us there are many more possible values the true function could take. This type of uncertainty is called *epistemic uncertainty*, which is the *reducible uncertainty* associated with lack of information. As we acquire more data, this type of uncertainty disappears, as there will be increasingly fewer solutions consistent with what we observe. Like with the posterior mean, we can compute the posterior variance (the variability of these functions in the posterior) in closed form. With shade, we show two times the posterior standard deviation on either side of the mean, creating a *credible interval* that has a 95% probability of containing the true value of the function for any input $x$.

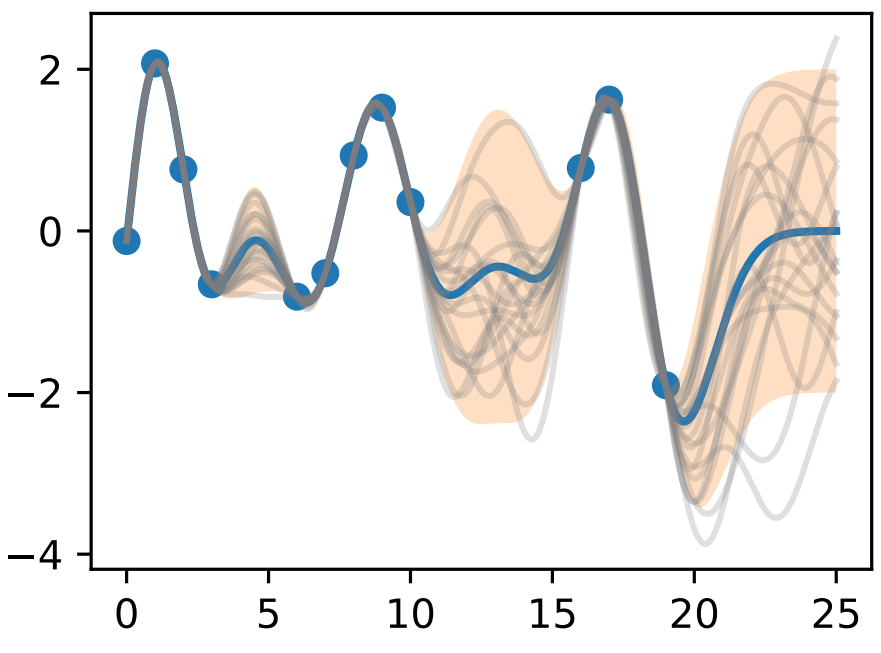

Fig. 18.1.5 Posterior samples, including 95% credible set.

The plot looks somewhat cleaner if we remove the posterior samples, simply visualizing the data, posterior mean, and 95% credible set. Notice how the uncertainty grows away from the data, a property of epistemic uncertainty.

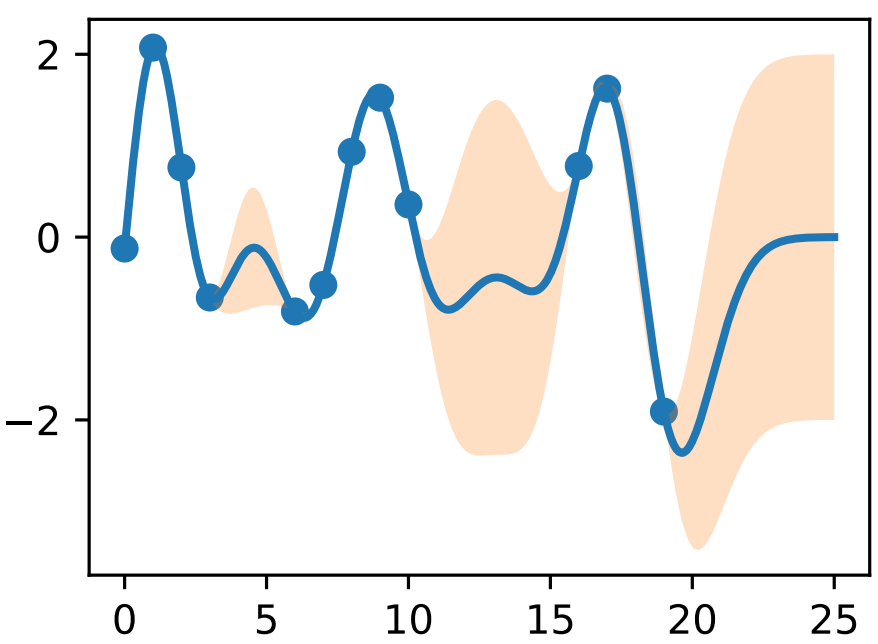

Fig. 18.1.6 Point predictions, and credible set.

The properties of the Gaussian process that we used to fit the data are strongly controlled by what's called a *covariance function*, also known as a *kernel*. The covariance function we used is called the *RBF (Radial Basis Function) kernel*, which has the form

$$k_{\text{RBF}}(x, x') = \text{Cov}(f(x), f(x')) = a^2 \exp\left(-\frac{1}{2\ell^2}||x - x'||^2\right) \tag{18.1.1}$$

The *hyperparameters* of this kernel are interpretable. The *amplitude* parameter $a$ controls the vertical scale over which the function is varying, and the *length-scale* parameter $\ell$ controls the rate of variation (the wiggliness) of the function. Larger $a$ means larger function values, and larger $\ell$ means more slowly varying functions. Let's see what happens to our sample prior and posterior functions as we vary $a$ and $\ell$.

The *length-scale* has a particularly pronounced effect on the predictions and uncertainty of a GP. At $||x - x'|| = \ell$ , the covariance between a pair of function values is $a^2 \exp(-0.5)$. At larger distances than $\ell$ , the values of the function values becomes nearly uncorrelated. This means that if we want to make a prediction at a point $x_*$, then function values with inputs $x$ such that $||x - x'|| > \ell$ will not have a strong effect on our predictions.

Let's see how changing the lengthscale affects sample prior and posterior functions, and credible sets. The above fits use a length-scale of $2$. Let's now consider $\ell = 0.1, 0.5, 2, 5, 10$ . A length-scale of $0.1$ is very small relative to the range of the input domain we are considering, $25$. For example, the values of the function at $x = 5$ and $x = 10$ will have essentially no correlation at such a length-scale. On the other hand, for a length-scale of $10$, the function values at these inputs will be highly correlated. Note that the vertical scale changes in the following figures.

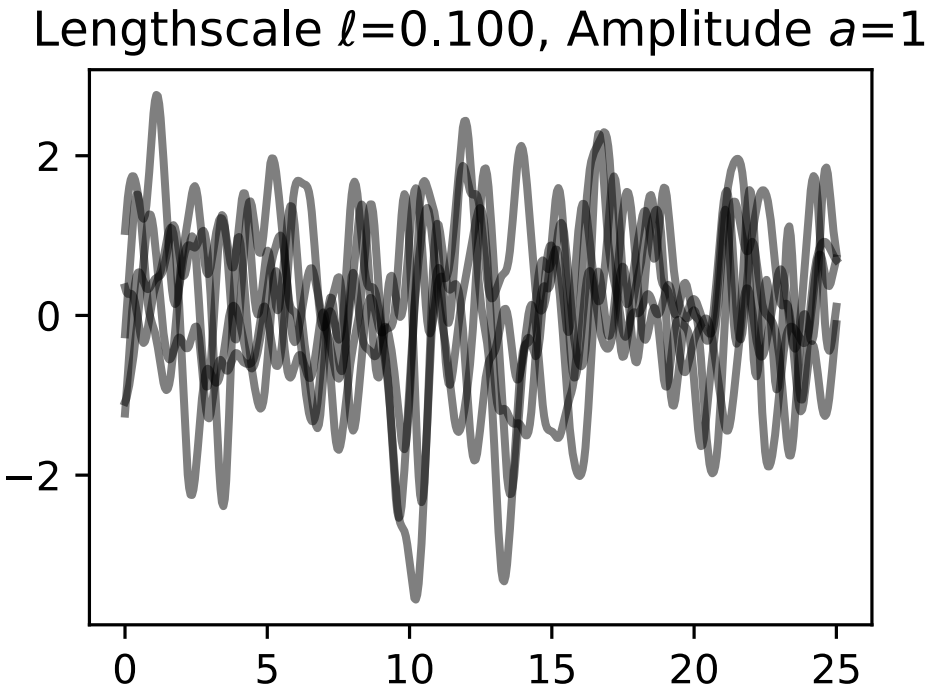

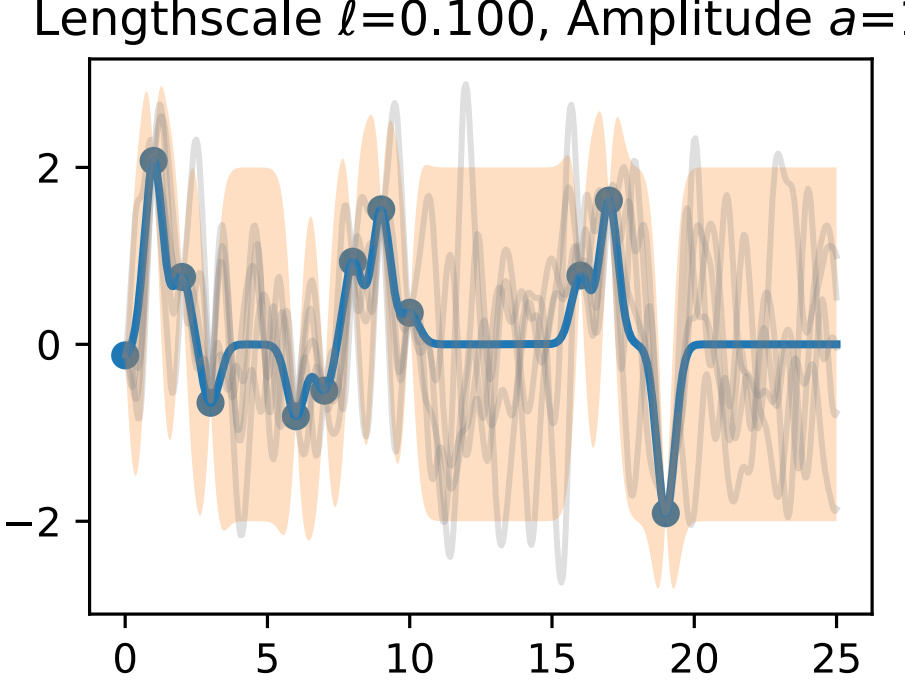

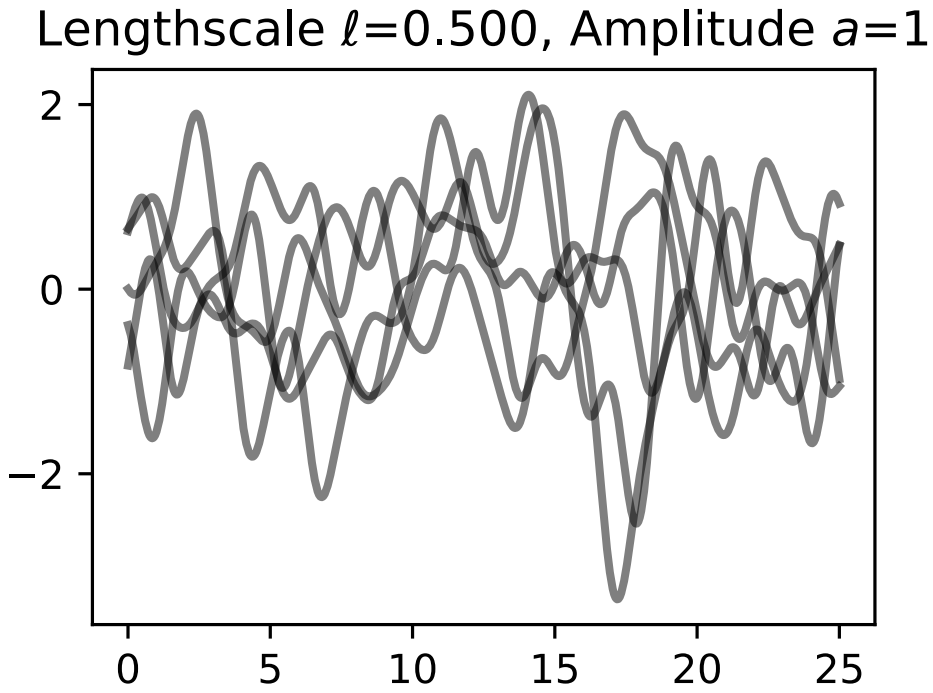

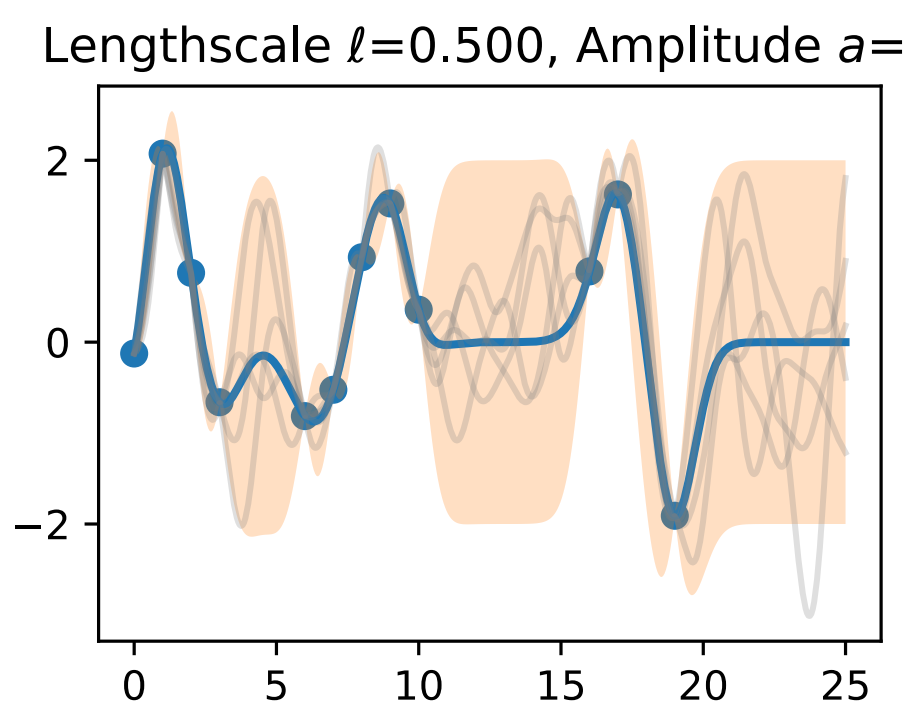

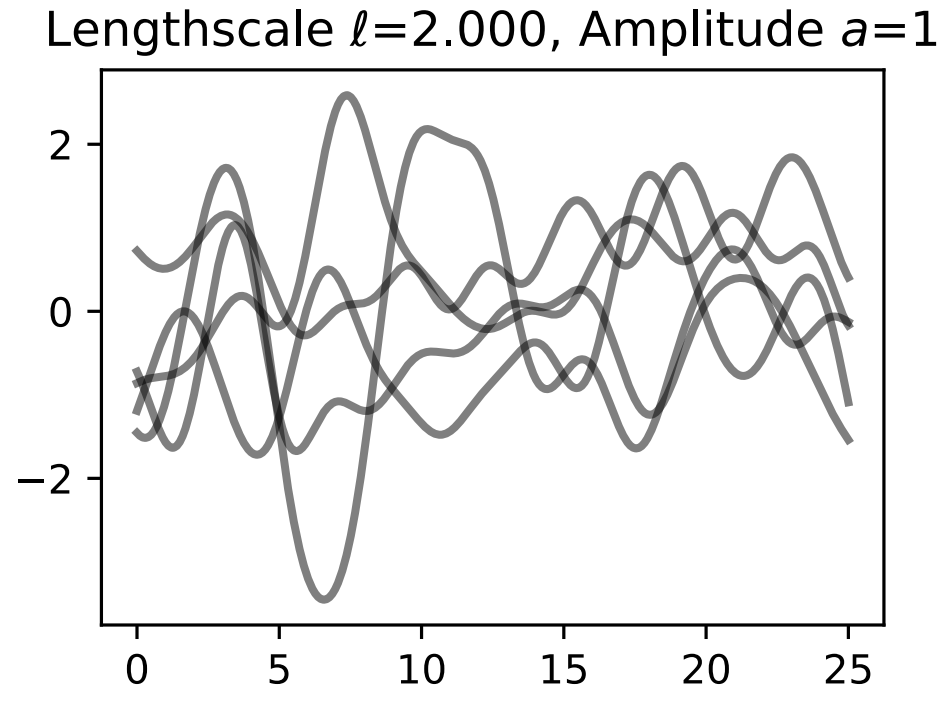

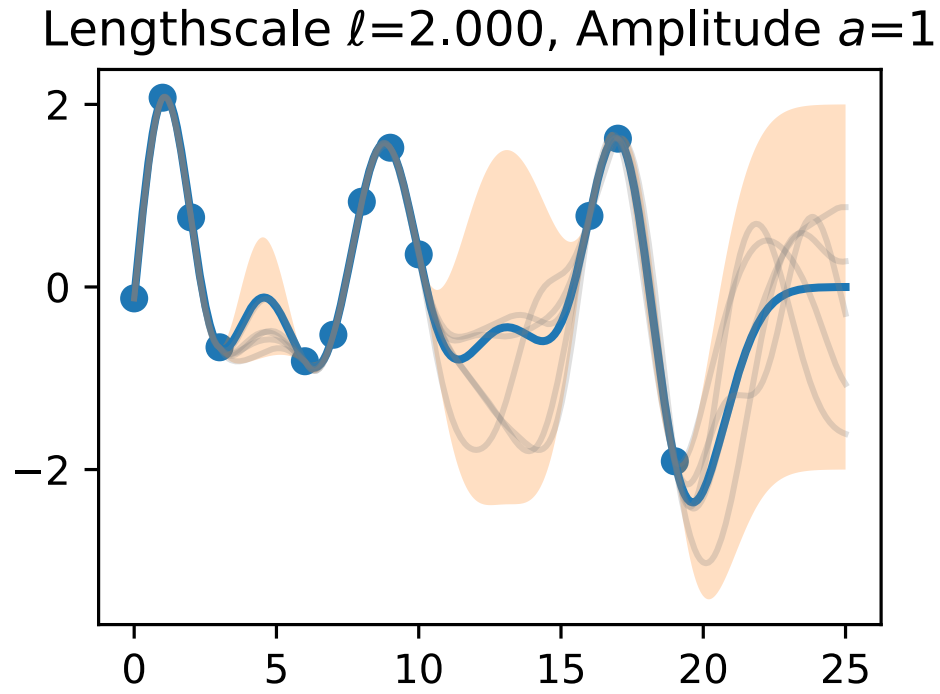

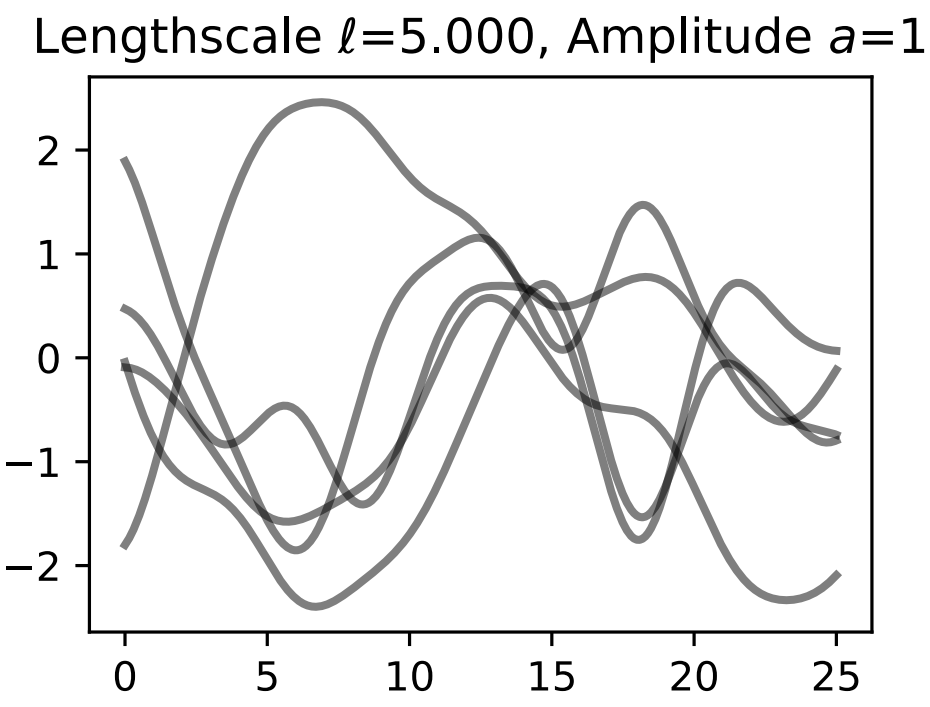

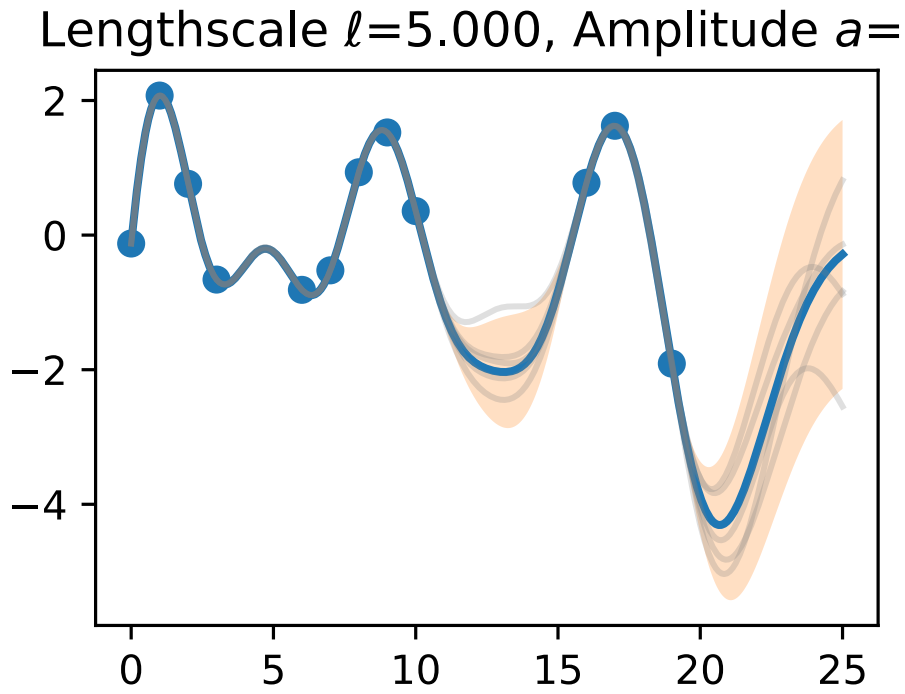

Notice as the length-scale increases the 'wiggliness' of the functions decrease, and our uncertainty decreases. If the length-scale is small, the uncertainty will quickly increase as we move away from the data, as the datapoints become less informative about the function values.

Now, let's vary the amplitude parameter, holding the length-scale fixed at 2. Note the vertical scale is held fixed for the prior samples, and varies for the posterior samples, so you can clearly see both the increasing scale of the function, and the fits to the data.

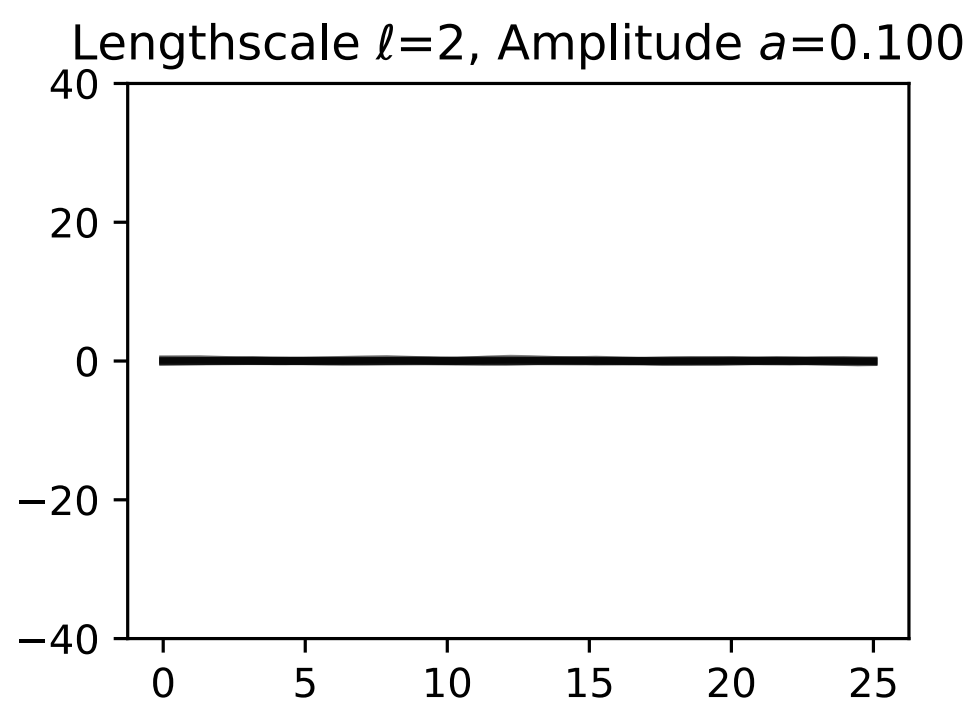

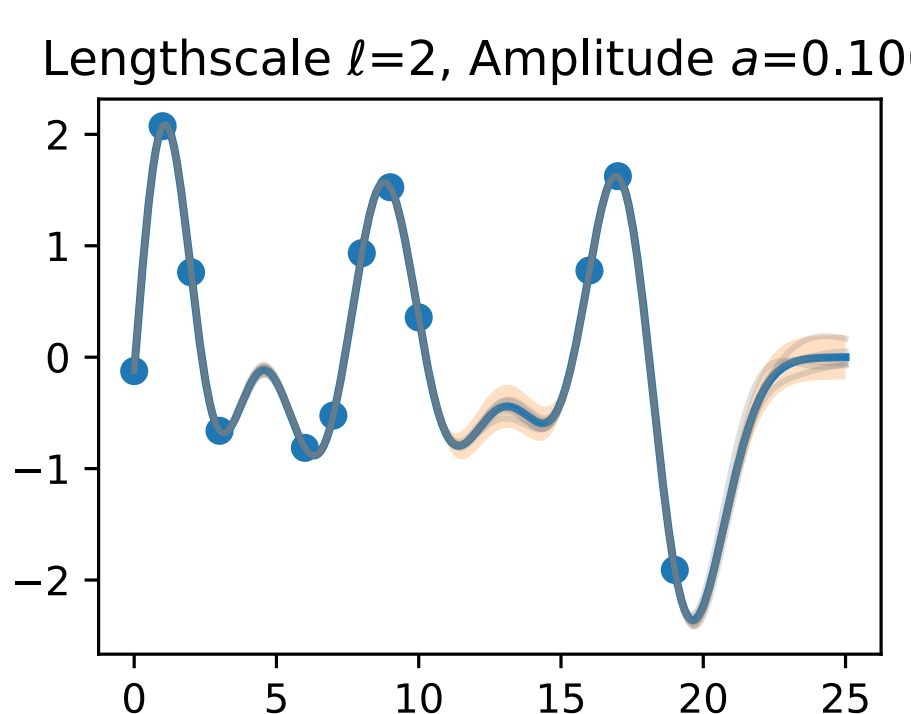

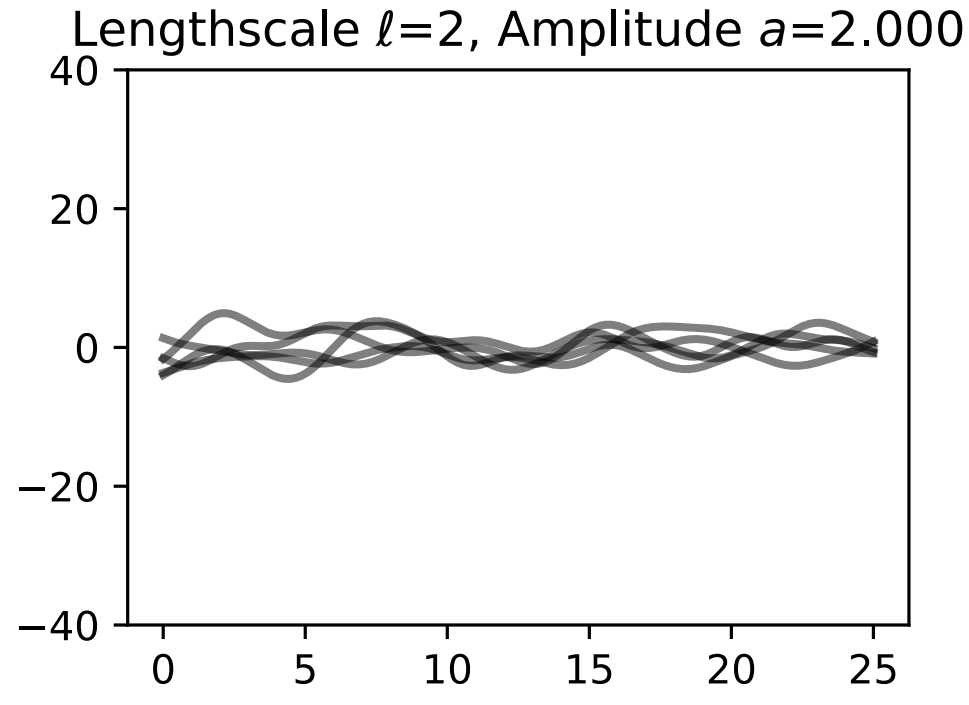

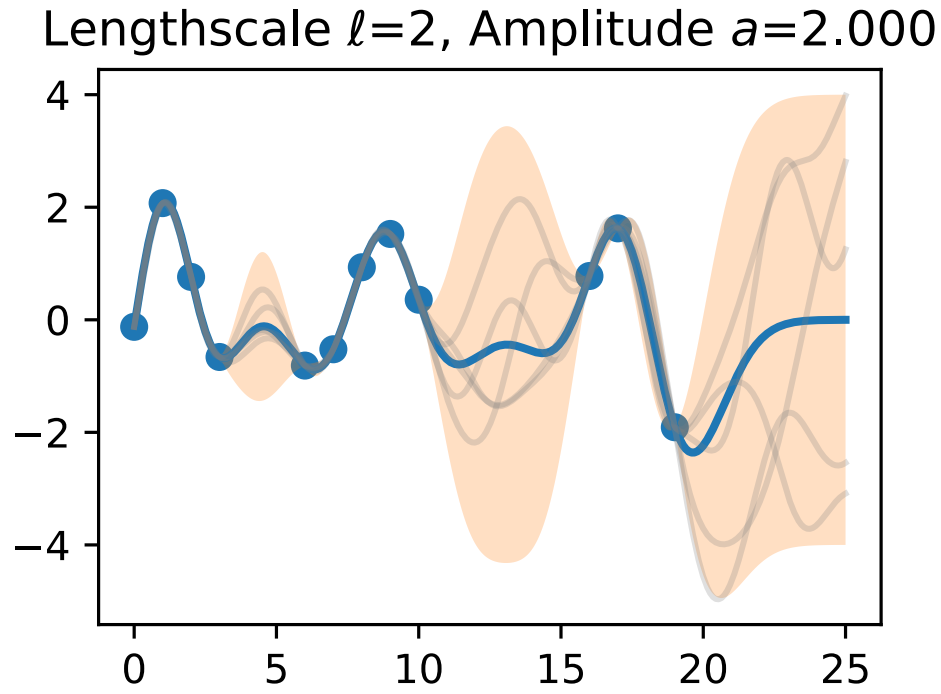

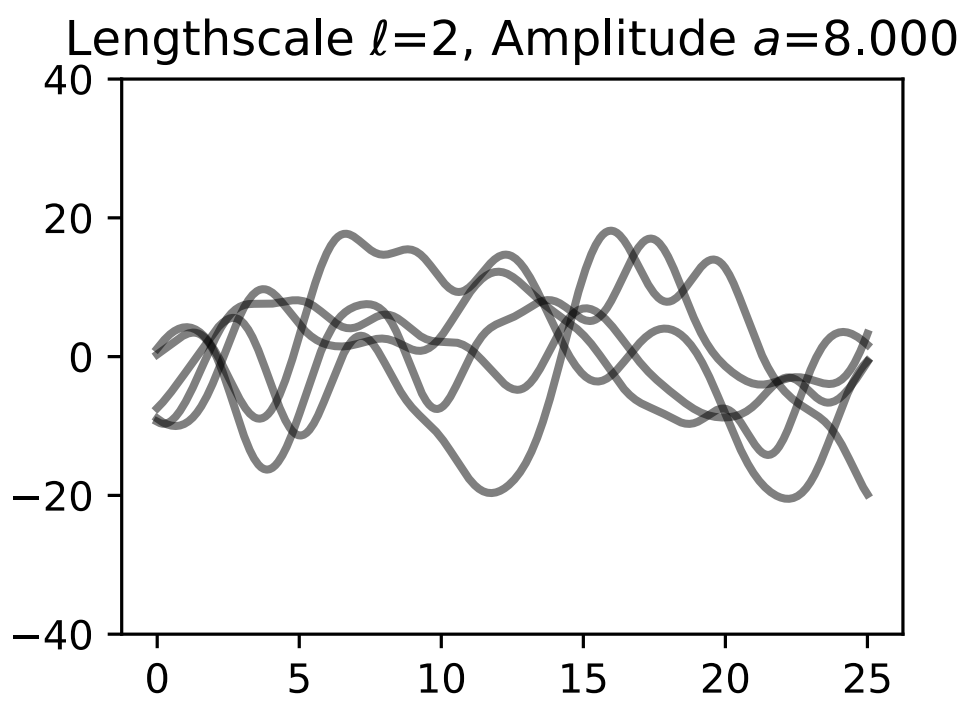

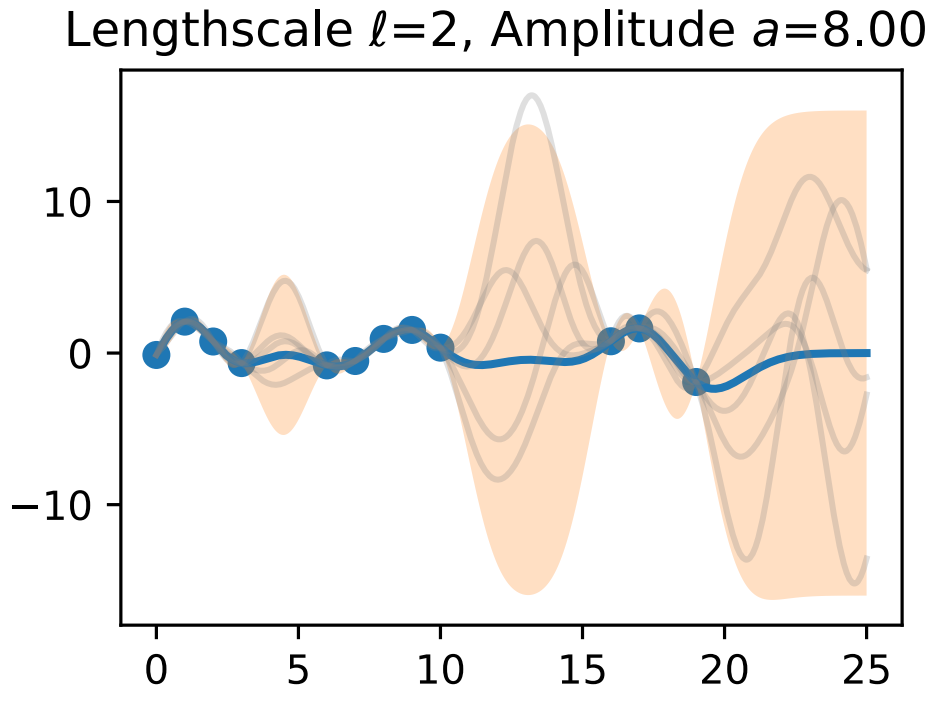

We see the amplitude parameter affects the scale of the function, but not the rate of variation. At this point, we also have the sense that the generalization performance of our procedure will depend on having reasonable values for these hyperparameters. Values of $\ell = 2$ and $a = 1$ appeared to provide reasonable fits, while some of the other values did not. Fortunately, there is a robust and automatic way to specify these hyperparameters, using what is called the *marginal likelihood*, which we will return to in the notebook on inference.

So what is a GP, really? As we started, a GP simply says that any collection of function values $f(x_1), \ldots, f(x_n)$, indexed by any collection of inputs $x_1, \ldots, x_n$ has a joint multivariate Gaussian distribution. The mean vector $\mu$ of this distribution is given by a *mean function*, which is typically taken to be a constant or zero. The covariance matrix of this distribution is given by the *kernel* evaluated at all pairs of the inputs $x$.

$$\begin{bmatrix} f(x) \\ f(x_1) \\ \vdots \\ f(x_n) \end{bmatrix} \sim \mathcal{N}\left(\mu, \begin{bmatrix} k(x,x) & k(x,x_1) & \dots & k(x,x_n) \\ k(x_1,x) & k(x_1,x_1) & \dots & k(x_1,x_n) \\ \vdots & \vdots & \ddots & \vdots \\ k(x_n,x) & k(x_n,x_1) & \dots & k(x_n,x_n) \end{bmatrix}\right) \tag{18.1.2}$$

Equation (18.1.2) specifies a GP prior. We can compute the conditional distribution of $f(x)$ for any $x$ given $f(x_1), \dots, f(x_n)$, the function values we have observed. This conditional distribution is called the *posterior*, and it is what we use to make predictions.

In particular,

$$f(x) | f(x_1), \dots, f(x_n) \sim \mathcal{N}(m, s^2) \tag{18.1.3}$$

where

$$m = k(x, x_{1:n}) k(x_{1:n}, x_{1:n})^{-1} f(x_{1:n}) \tag{18.1.4}$$

$$s^2 = k(x,x) - k(x, x_{1:n}) k(x_{1:n}, x_{1:n})^{-1} k(x, x_{1:n}) \tag{18.1.5}$$

where $k(x, x_{1:n})$ is a $1 \times n$ vector formed by evaluating $k(x, x_i)$ for $i = 1, \dots, n$ and $k(x_{1:n}, x_{1:n})$ is an $n \times n$ matrix formed by evaluating $k(x_i, x_j)$ for $i, j = 1, \dots, n$. $m$ is what we can use as a point predictor for any $x$, and $s^2$ is what we use for uncertainty: if we want to create an interval with a 95% probability that $f(x)$ is in the interval, we would use $m \pm 2s$. The predictive means and uncertainties for all the above figures were created using these equations. The observed data points were given by $f(x_1), \dots, f(x_n)$ and chose a fine grained set of $x$ points to make predictions.

Let's suppose we observe a single datapoint, $f(x_1)$, and we want to determine the value of $f(x)$ at some $x$. Because $f(x)$ is described by a Gaussian process, we know the joint distribution over $(f(x), f(x_1))$ is Gaussian:

$$\begin{bmatrix} f(x) \\ f(x_1) \end{bmatrix} \sim \mathcal{N}\left(\mu, \begin{bmatrix} k(x,x) & k(x,x_1) \\ k(x_1,x) & k(x_1,x_1) \end{bmatrix}\right) \tag{18.1.6}$$

The off-diagonal expression $k(x, x_1) = k(x_1, x)$ tells us how correlated the function values will be — how strongly determined $f(x)$ will be from $f(x_1)$. We have seen already that if we use a large length-scale, relative to the distance between $x$ and $x_1$, $||x - x_1||$, then the function values will be highly correlated. We can visualize the process of determining $f(x)$ from $f(x_1)$ both in the space of functions, and in the joint distribution over $f(x_1), f(x)$. Let's initially consider an $x$ such that $k(x, x_1) = 0.9$, and $k(x, x) = 1$, meaning that the value of $f(x)$ is moderately correlated with the value of $f(x_1)$. In the joint distribution, the contours of constant probability will be relatively narrow ellipses.

Suppose we observe $f(x_1) = 1.2$. To condition on this value of $f(x_1)$, we can draw a horizontal line at $1.2$ on our plot of the density, and see that the value of $f(x)$ is mostly constrained to $[0.64, 1.52]$. We have also drawn this plot in function space, showing the observed point $f(x_1)$ in orange, and 1 standard deviation of the Gaussian process predictive distribution for $f(x)$ in blue, about the mean value of $1.08$.

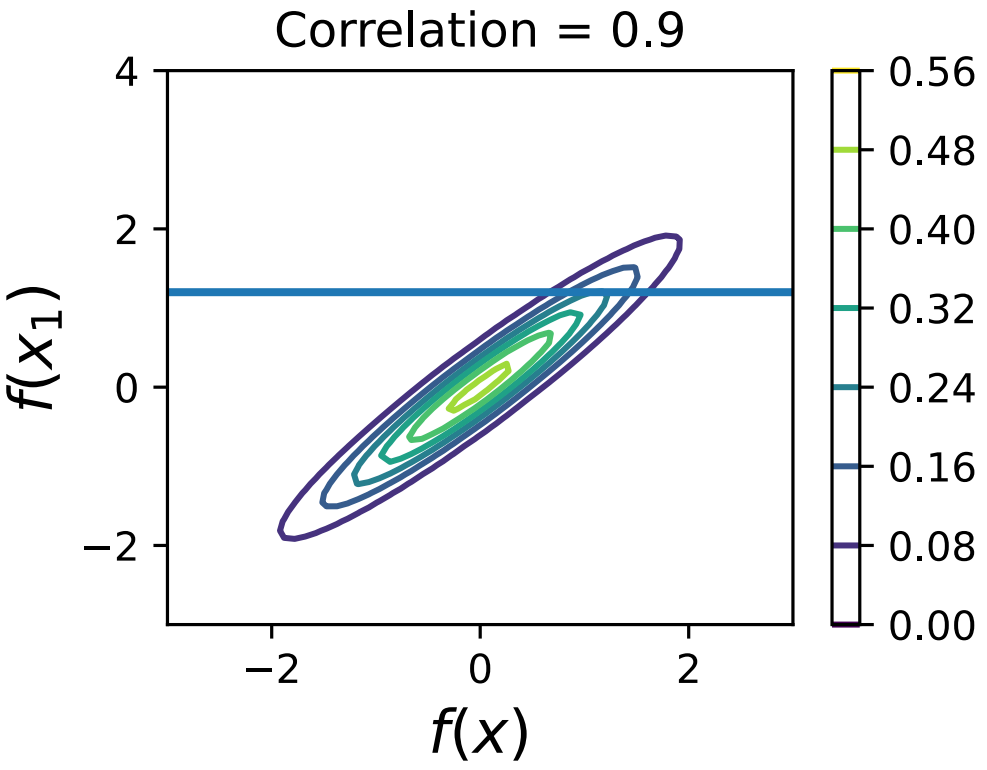

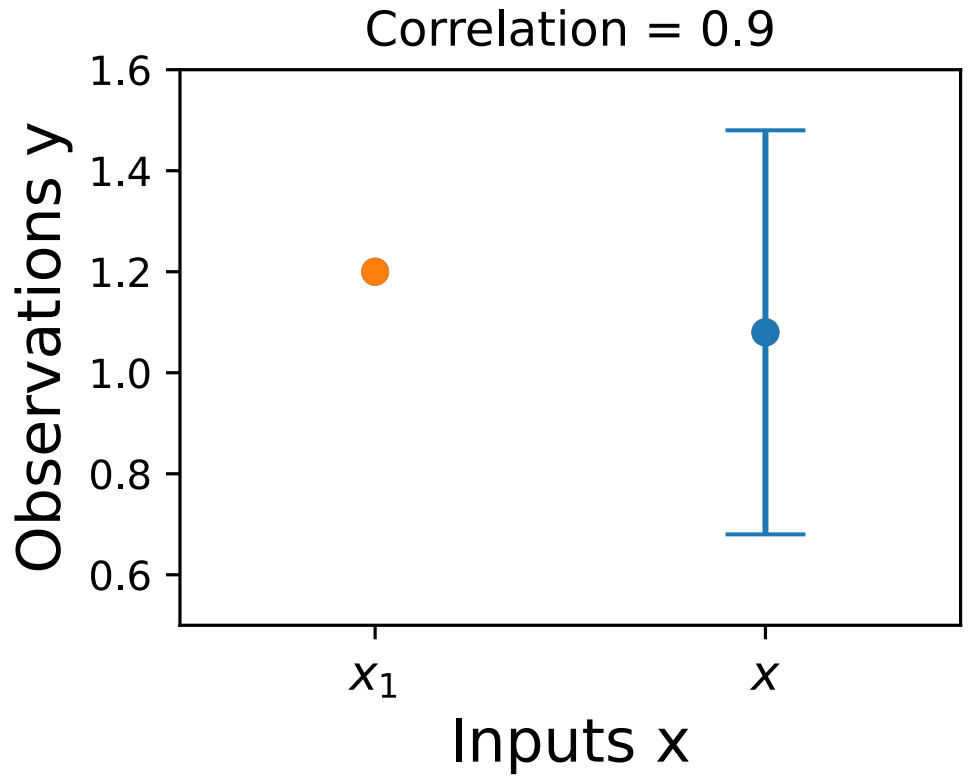

Now suppose we have a stronger correlation, $k(x, x_1) = 0.95$. Now the ellipses have narrowed further, and the value of $f(x)$ is even more strongly determined by $f(x_1)$. Drawing a horizontal line at $1.2$, we see the contours for $f(x)$ support values mostly within $[0.83, 1.45]$. Again, we also show the plot in function space, with one standard deviation about the mean predictive value of $1.14$.

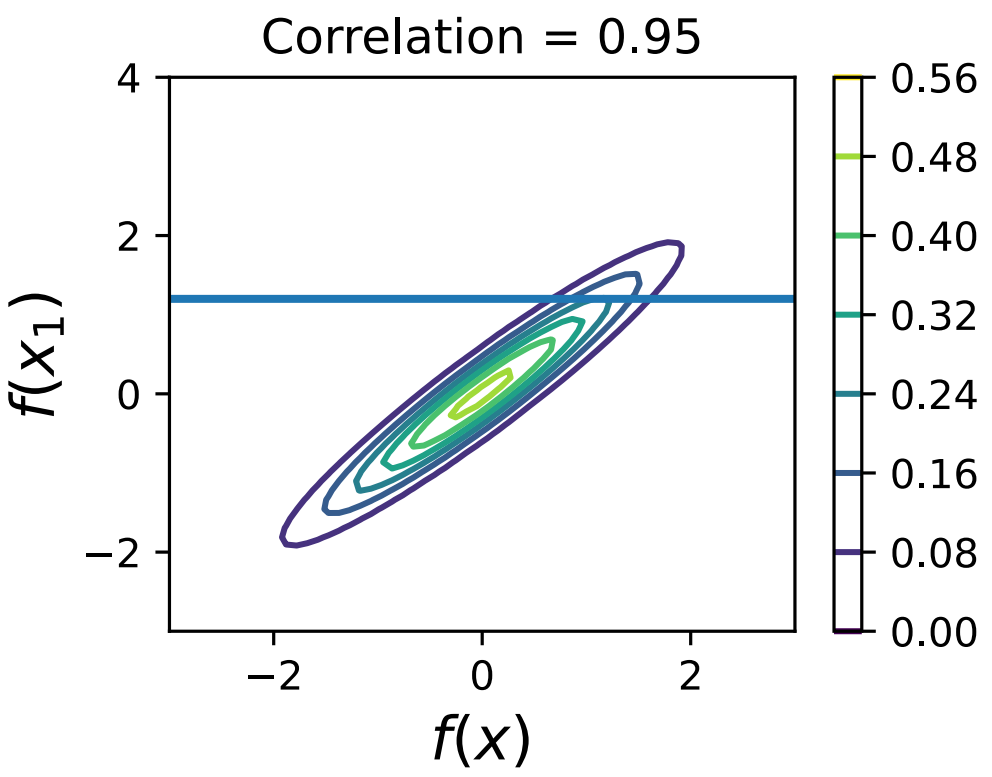

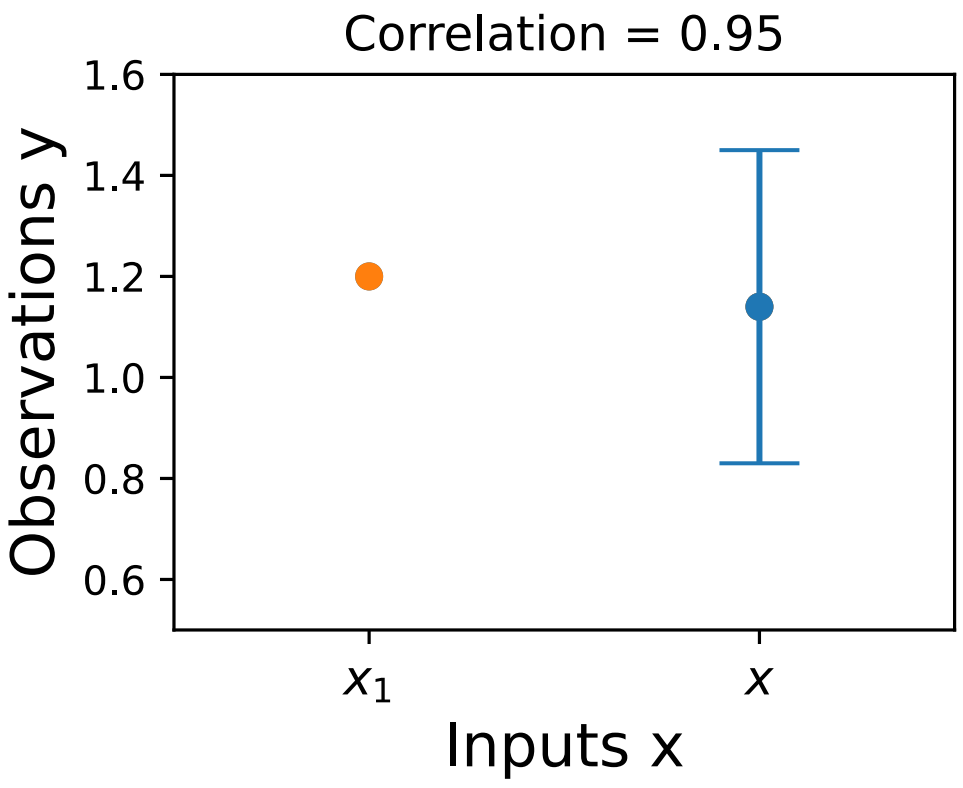

We see that the posterior mean predictor of our Gaussian process is closer to 1.2, because there is now a stronger correlation. We also see that our uncertainty (the error bars) have somewhat decreased. Despite the strong correlation between these function values, our uncertainty is still righly quite large, because we have only observed a single data point!

This procedure can give us a posterior on $f(x)$ for any $x$, for any number of points we have observed. Suppose we observe $f(x_1), f(x_2)$. We now visualize the posterior for $f(x)$ at a particular $x = x'$ in function space. The exact distribution for $f(x)$ is given by the above equations. $f(x)$ is Gaussian distributed, with mean

$$m = k(x, x_{1:3})k(x_{1:3}, x_{1:3})^{-1} f(x_{1:3}) \tag{18.1.7}$$

and variance

$$s^2 = k(x, x) - k(x, x_{1:3})k(x_{1:3}, x_{1:3})^{-1} k(x, x_{1:3}) \tag{18.1.8}$$

In this introductory notebook, we have been considering *noise free* observations. As we will see, it is easy to include observation noise. If we assume that the data are generated from a latent noise free function $f(x)$ plus iid Gaussian noise $\epsilon(x) \sim \mathcal{N}(0, \sigma^2)$ with variance $\sigma^2$, then our covariance function simply becomes $k(x_i, x_j) \to k(x_i, x_j) + \delta_{ij}\sigma^2$, where $\delta_{ij} = 1$ if $i = j$ and 0 otherwise.

We have already started getting some intuition about how we can use a Gaussian process to specify a prior and posterior over solutions, and how the kernel function affects the properties of these solutions. In the following notebooks, we will precisely show how to specify a Gaussian process prior, introduce and derive various kernel functions, and then go through the mechanics of how to automatically learn kernel hyperparameters, and form a Gaussian process posterior to make predictions. While it takes time and practice to get used to concepts such as a “distributions over functions”, the actual mechanics of finding the GP predictive equations is actually quite simple — making it easy to get practice to form an intuitive understanding of these concepts.

## 18.1.1 Summary

In typical machine learning, we specify a function with some free parameters (such as a neural network and its weights), and we focus on estimating those parameters, which may not be interpretable. With a Gaussian process, we instead reason about distributions over functions directly, which enables us to reason about the high-level properties of the solutions. These properties are controlled by a covariance function (kernel), which often has a few highly interpretable hyperparameters. These hyperparameters include the *length-scale*, which controls how rapidly (how wiggily) the functions are. Another hyperparameter is the amplitude, which controls the vertical scale over which our functions are varying. Representing many different functions that can fit the data, and combining them all together into a predictive distribution, is a distinctive feature of Bayesian methods. Because there is a greater amount of variability between possible solutions far away from the data, our uncertainty intuitively grows as we move from the data.

A Gaussian process represents a distribution over functions by specifying a multivariate normal (Gaussian) distribution over all possible function values. It is possible to easily manipulate Gaussian distributions to find the distribution of one function value based on the values of any set of other values. In other words, if we observe a set of points, then we can condition on these points and infer a distribution over what the value of the function might look like at any other input. How we model the correlations between these points is determined by the covariance function and is what defines the generalization properties of the Gaussian process. While it takes time to get used to Gaussian processes, they are easy to work with, have many applications, and help us understand and develop other model classes, like neural networks.

## 18.1.2 Exercises

1. What is the difference between epistemic uncertainty versus observation uncertainty?
2. Besides rate of variation and amplitude, what other properties of functions might we want to consider, and what would be real-world examples of functions that have those properties?

3. The RBF covariance function we considered says that covariances (and correlations) between observations decrease with their distance in the input space (times, spatial locations, etc.). Is this a reasonable assumption? Why or why not?
4. Is a sum of two Gaussian variables Gaussian? Is a product of two Gaussian variables Gaussian? If (a,b) have a joint Gaussian distribution, is a|b (a given b) Gaussian? Is a Gaussian?
5. Repeat the exercise where we observe a data point at $f(x_1) = 1.2$, but now suppose we additionally observe $f(x_2) = 1.4$. Let $k(x, x_1) = 0.9$, and $k(x, x_2) = 0.8$. Will we be more or less certain about the value of $f(x)$, than when we had only observed $f(x_1)$? What is the mean and 95% credible set for our value of $f(x)$ now?
6. Do you think increasing our estimate of observation noise would increase or decrease our estimate of the length-scale of the ground truth function?
7. As we move away from the data, suppose the uncertainty in our predictive distribution increases to a point, then stops increasing. Why might that happen?

Discussions[259].

## 18.2 Gaussian Process Priors

Understanding Gaussian processes (GPs) is important for reasoning about model construction and generalization, and for achieving state-of-the-art performance in a variety of applications, including active learning, and hyperparameter tuning in deep learning. GPs are everywhere, and it is in our interests to know what they are and how we can use them.

In this section, we introduce Gaussian process *priors* over functions. In the next notebook, we show how to use these priors to do *posterior inference* and make predictions. The next section can be viewed as "GPs in a nutshell", quickly giving what you need to apply Gaussian processes in practice.

```
import numpy as np
from scipy.spatial import distance_matrix
from d2l import torch as d2l

d2l.set_figsize()
```

### 18.2.1 Definition

A Gaussian process is defined as *a collection of random variables, any finite number of which have a joint Gaussian distribution*. If a function $f(x)$ is a Gaussian process, with *mean function* $m(x)$ and *covariance function* or *kernel* $k(x, x')$, $f(x) \sim \mathcal{GP}(m, k)$, then any collec-tion of function values queried at any collection of input points $x$ (times, spatial locations, image pixels, etc.), has a joint multivariate Gaussian distribution with mean vector $\mu$ and covariance matrix $K$: $f(x_1), \dots, f(x_n) \sim \mathcal{N}(\mu, K)$, where $\mu_i = E[f(x_i)] = m(x_i)$ and $K_{ij} = \mathrm{Cov}(f(x_i), f(x_j)) = k(x_i, x_j)$.

This definition may seem abstract and inaccessible, but Gaussian processes are in fact very simple objects. Any function

$$f(x) = w^\top \phi(x) = \langle w, \phi(x) \rangle, \tag{18.2.1}$$

with $w$ drawn from a Gaussian (normal) distribution, and $\phi$ being any vector of basis functions, for example $\phi(x) = (1, x, x^2, ..., x^d)^\top$, is a Gaussian process. Moreover, any Gaussian process f(x) can be expressed in the form of equation (18.2.1). Let's consider a few concrete examples, to begin getting acquainted with Gaussian processes, after which we can appreciate how simple and useful they really are.

## 18.2.2 A Simple Gaussian Process

Suppose $f(x) = w_0 + w_1 x$, and $w_0, w_1 \sim \mathcal{N}(0, 1)$, with $w_0, w_1, x$ all in one dimension. We can equivalently write this function as the inner product $f(x) = (w_0, w_1)(1, x)^\top$. In (18.2.1) above, $w = (w_0, w_1)^\top$ and $\phi(x) = (1, x)^\top$.

For any $x$, $f(x)$ is a sum of two Gaussian random variables. Since Gaussians are closed under addition, $f(x)$ is also a Gaussian random variable for any $x$. In fact, we can compute for any particular $x$ that $f(x)$ is $\mathcal{N}(0, 1 + x^2)$. Similarly, the joint distribution for any collection of function values, $(f(x_1), \dots, f(x_n))$, for any collection of inputs $x_1, \dots, x_n$, is a multivariate Gaussian distribution. Therefore $f(x)$ is a Gaussian process.

In short, $f(x)$ is a *random function*, or a *distribution over functions*. We can gain some insights into this distribution by repeatedly sampling values for $w_0, w_1$, and visualizing the corresponding functions $f(x)$, which are straight lines with slopes and different intercepts, as follows:

```
def lin_func(x, n_sample):
    preds = np.zeros((n_sample, x.shape[0]))
    for ii in range(n_sample):
        w = np.random.normal(0, 1, 2)
        y = w[0] + w[1] * x
        preds[ii, :] = y
    return preds

x_points = np.linspace(-5, 5, 50)
outs = lin_func(x_points, 10)
lw_bd = -2 * np.sqrt((1 + x_points ** 2))
up_bd = 2 * np.sqrt((1 + x_points ** 2))

d2l.plt.fill_between(x_points, lw_bd, up_bd, alpha=0.25)
d2l.plt.plot(x_points, np.zeros(len(x_points)), linewidth=4, color='black')
d2l.plt.plot(x_points, outs.T)
```

```
d2l.plt.xlabel("x", fontsize=20)
d2l.plt.ylabel("f(x)", fontsize=20)
d2l.plt.show()
```

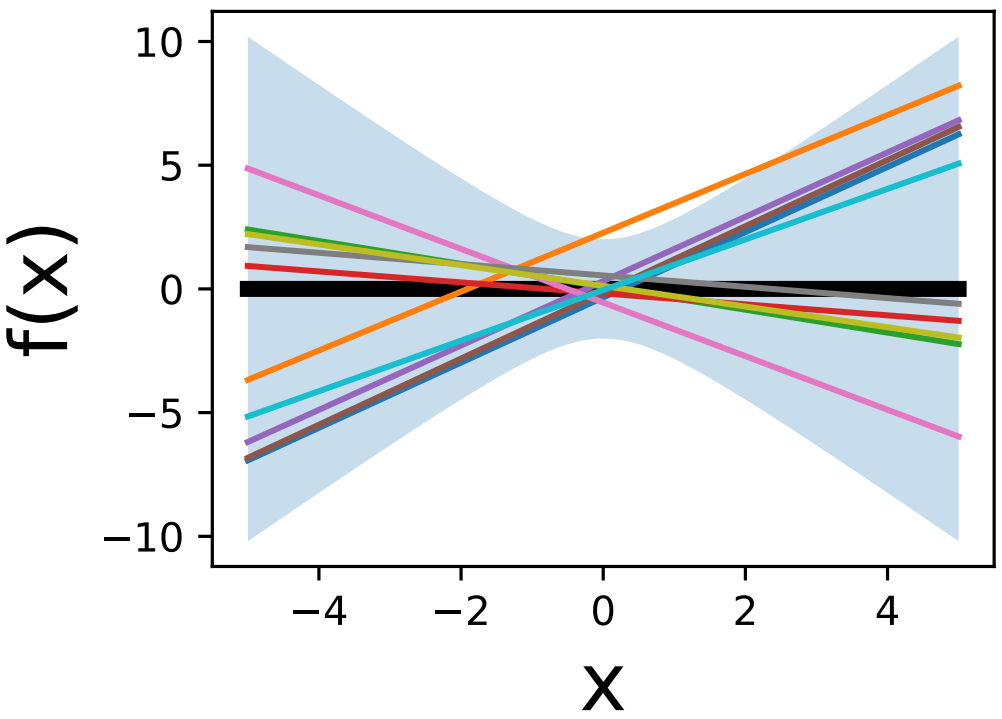

If $w_0$ and $w_1$ are instead drawn from $\mathcal{N}(0, \alpha^2)$, how do you imagine varying $\alpha$ affects the distribution over functions?

### 18.2.3 From Weight Space to Function Space

In the plot above, we saw how a distribution over parameters in a model induces a distribution over functions. While we often have ideas about the functions we want to model — whether they're smooth, periodic, quickly varying, etc. — it is relatively tedious to reason about the parameters, which are largely uninterpretable. Fortunately, Gaussian processes provide an easy mechanism to reason *directly* about functions. Since a Gaussian distribution is entirely defined by its first two moments, its mean and covariance matrix, a Gaussian process by extension is defined by its mean function and covariance function.

In the above example, the mean function

$$m(x) = E[f(x)] = E[w_0 + w_1 x] = E[w_0] + E[w_1]x = 0 + 0 = 0. \tag{18.2.2}$$

Similarly, the covariance function is

$$k(x, x') = \mathrm{Cov}(f(x), f(x')) = E[f(x)f(x')] - E[f(x)]E[f(x')] = E[w_0^2 + w_0 w_1 x' + w_1 w_0 x + w_1^2 x x'] = 1 + xx \tag{18.2.3}$$

Our distribution over functions can now be directly specified and sampled from, without needing to sample from the distribution over parameters. For example, to draw from $f(x)$, we can simply form our multivariate Gaussian distribution associated with any collection of $x$ we want to query, and sample from it directly. We will begin to see just how advantageous this formulation will be.

First, we note that essentially the same derivation for the simple straight line model above can be applied to find the mean and covariance function for *any* model of the form $f(x) = w^\top \phi(x)$, with $w \sim \mathcal{N}(u, S)$. In this case, the mean function $m(x) = u^\top \phi(x)$, and the covariance function $k(x, x') = \phi(x)^\top S \phi(x')$. Since $\phi(x)$ can represent a vector of any non-linear basis functions, we are considering a very general model class, including models with an even an *infinite* number of parameters.

### 18.2.4 The Radial Basis Function (RBF) Kernel

The *radial basis function* (RBF) kernel is the most popular covariance function for Gaussian processes, and kernel machines in general. This kernel has the form $k_{\textrm{RBF}}(x, x') = a^2 \exp\left(-\frac{1}{2\ell^2}||x - x'||^2\right)$, where $a$ is an amplitude parameter, and $\ell$ is a *lengthscale* hyperparameter.

Let's derive this kernel starting from weight space. Consider the function

$$f(x) = \sum_{i=1}^{J} w_i \phi_i(x), w_i \sim \mathcal{N}\left(0, \frac{\sigma^2}{J}\right), \phi_i(x) = \exp\left(-\frac{(x - c_i)^2}{2\ell^2}\right). \tag{18.2.4}$$

$f(x)$ is a sum of radial basis functions, with width $\ell$, centred at the points $c_i$, as shown in the following figure.

We can recognize $f(x)$ as having the form $w^\top \phi(x)$, where $w = (w_1, \dots, w_J)^\top$ and $\phi(x)$ is a vector containing each of the radial basis functions. The covariance function of this Gaussian process is then

$$k(x, x') = \frac{\sigma^2}{J} \sum_{i=1}^{J} \phi_i(x) \phi_i(x'). \tag{18.2.5}$$

Now let's consider what happens as we take the number of parameters (and basis functions) to infinity. Let $c_J = \log J$, $c_1 = -\log J$, and $c_{i+1} - c_i = \Delta c = 2\frac{\log J}{J}$, and $J \to \infty$. The covariance function becomes the Riemann sum:

$$k(x, x') = \lim_{J \to \infty} \frac{\sigma^2}{J} \sum_{i=1}^{J} \phi_i(x) \phi_i(x') = \int_{c_0}^{c_\infty} \phi_c(x) \phi_c(x') dc. \tag{18.2.6}$$

By setting $c_0 = -\infty$ and $c_\infty = \infty$, we spread the infinitely many basis functions across the whole real line, each a distance $\Delta c \to 0$ apart:

$$k(x, x') = \int_{-\infty}^{\infty} \exp(-\frac{(x - c)^2}{2\ell^2}) \exp(-\frac{(x' - c)^2}{2\ell^2}) dc = \sqrt{\pi} \ell \sigma^2 \exp(-\frac{(x - x')^2}{2(\sqrt{2}\ell)^2}) \propto k_{\textrm{RBF}}(x, x'). \tag{18.2.7}$$

It is worth taking a moment to absorb what we have done here. By moving into the function space representation, we have derived how to represent a model with an *infinite* number of parameters, using a finite amount of computation. A Gaussian process with an RBF kernel is a *universal approximator*, capable of representing any continuous function to arbitrary precision. We can intuitively see why from the above derivation. We can collapse each radial basis function to a point mass taking $\ell \to 0$, and give each point mass any height we wish.

So a Gaussian process with an RBF kernel is a model with an infinite number of parameters and much more flexibility than any finite neural network. Perhaps all the fuss about *overparametrized* neural networks is misplaced. As we will see, GPs with RBF kernels do not overfit, and in fact provide especially compelling generalization performance on small datasets. Moreover, the examples in (Zhang *et al.*, 2021), such as the ability to fit images with random labels perfectly, but still generalize well on structured problems, (can be perfectly reproduced using Gaussian processes) (Wilson and Izmailov, 2020). Neural networks are not as distinct as we make them out to be.

We can build further intuition about Gaussian processes with RBF kernels, and hyperparameters such as *length-scale*, by sampling directly from the distribution over functions. As before, this involves a simple procedure:

1. Choose the input $x$ points we want to query the GP: $x_1, \dots, x_n$.
2. Evaluate $m(x_i)$, $i = 1, \dots, n$, and $k(x_i, x_j)$ for $i, j = 1, \dots, n$ to respectively form the mean vector and covariance matrix $\mu$ and $K$, where $(f(x_1), \dots, f(x_n)) \sim \mathcal{N}(\mu, K)$.
3. Sample from this multivariate Gaussian distribution to obtain the sample function values.
4. Sample more times to visualize more sample functions queried at those points.

We illustrate this process in the figure below.

```
def rbfkernel(x1, x2, ls=4.):  #@save
    dist = distance_matrix(np.expand_dims(x1, 1), np.expand_dims(x2, 1))
    return np.exp(-(1. / ls / 2) * (dist ** 2))

x_points = np.linspace(0, 5, 50)
meanvec = np.zeros(len(x_points))
covmat = rbfkernel(x_points,x_points, 1)

prior_samples= np.random.multivariate_normal(meanvec, covmat, size=5);
d2l.plt.plot(x_points, prior_samples.T, alpha=0.5)
d2l.plt.show()
```

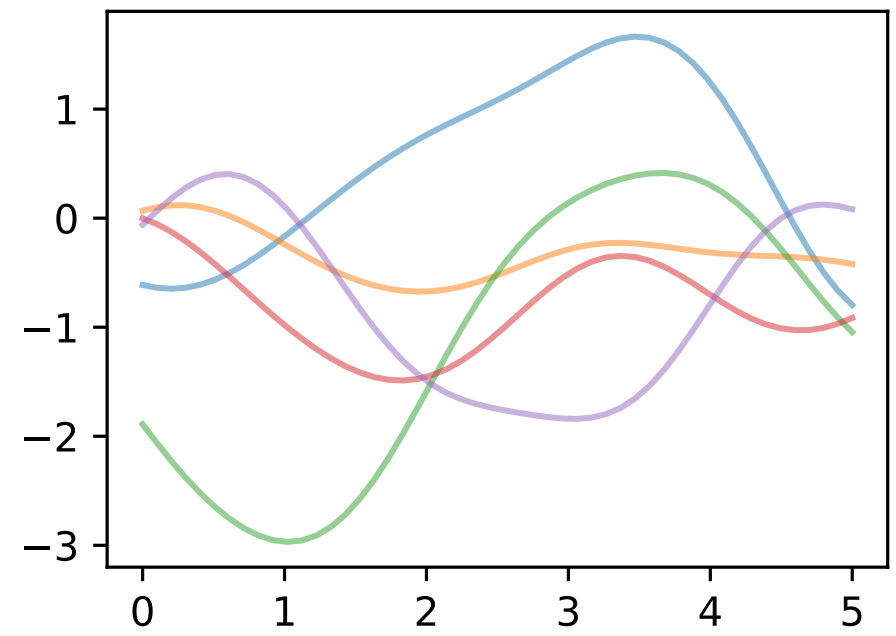

## 18.2.5 The Neural Network Kernel

Research on Gaussian processes in machine learning was triggered by research on neural networks. Radford Neal was pursuing ever larger Bayesian neural networks, ultimately showing in 1994 (later published in 1996, as it was one of the most infamous NeurIPS rejections) that such networks with an infinite number of hidden units become Gaussian processes with particular kernel functions (Neal, 1996). Interest in this derivation has re-surfaced, with ideas like the neural tangent kernel being used to investigate the generalization properties of neural networks (Matthews *et al.*, 2018) (Novak *et al.*, 2018). We can derive the neural network kernel as follows.

Consider a neural network function $f(x)$ with one hidden layer:

$$f(x) = b + \sum_{i=1}^{J} v_i h(x; u_i). \tag{18.2.8}$$

$b$ is a bias, $v_i$ are the hidden to output weights, $h$ is any bounded hidden unit transfer function, $u_i$ are the input to hidden weights, and $J$ is the number of hidden units. Let $b$ and $v_i$ be independent with zero mean and variances $\sigma_b^2$ and $\sigma_v^2/J$, respectively, and let the $u_i$ have independent identical distributions. We can then use the central limit theorem to show that any collection of function values $f(x_1), \ldots, f(x_n)$ has a joint multivariate Gaussian distribution.

The mean and covariance function of the corresponding Gaussian process are:

$$m(x) = E[f(x)] = 0 \tag{18.2.9}$$

$$k(x, x') = \text{cov}[f(x), f(x')] = E[f(x)f(x')] = \sigma_b^2 + \frac{1}{J}\sum_{i=1}^{J} \sigma_v^2 E[h_i(x; u_i)h_i(x'; u_i)] \tag{18.2.10}$$

In some cases, we can essentially evaluate this covariance function in closed form. Let $h(x; u) = \text{erf}(u_0 + \sum_{j=1}^{P} u_j x_j)$, where $\text{erf}(z) = \frac{2}{\sqrt{\pi}} \int_0^z e^{-t^2} dt$, and $u \sim \mathcal{N}(0, \Sigma)$. Then $k(x, x') = \frac{2}{\pi} \sin(\frac{2\tilde{x}^\top \Sigma \tilde{x}'}{\sqrt{(1+2\tilde{x}^\top \Sigma \tilde{x})(1+2\tilde{x}'^\top \Sigma \tilde{x}')}})$.

The RBF kernel is *stationary*, meaning that it is *translation invariant*, and therefore can be written as a function of $\tau = x - x'$. Intuitively, stationarity means that the high-level properties of the function, such as rate of variation, do not change as we move in input space. The neural network kernel, however, is *non-stationary*. Below, we show sample functions from a Gaussian process with this kernel. We can see that the function looks qualitatively different near the origin.

### 18.2.6 Summary

The first step in performing Bayesian inference involves specifying a prior. Gaussian processes can be used to specify a whole prior over functions. Starting from a traditional "weight space" view of modelling, we can induce a prior over functions by starting with the functional form of a model, and introducing a distribution over its parameters. We can alternatively specify a prior distribution directly in function space, with properties controlled by a kernel. The function-space approach has many advantages. We can build models that actually correspond to an infinite number of parameters, but use a finite amount of computation! Moreover, while these models have a great amount of flexibility, they also make strong assumptions about what types of functions are a priori likely, leading to relatively good generalization on small datasets.

The assumptions of models in function space are intuitively controlled by kernels, which often encode higher level properties of functions, such as smoothness and periodicity. Many kernels are stationary, meaning that they are translation invariant. Functions drawn from a Gaussian process with a stationary kernel have roughly the same high-level properties (such as rate of variation) regardless of where we look in the input space.

Gaussian processes are a relatively general model class, containing many examples of models we are already familiar with, including polynomials, Fourier series, and so on, as long as we have a Gaussian prior over the parameters. They also include neural networks with an infinite number of parameters, even without Gaussian distributions over the parameters. This connection, discovered by Radford Neal, triggered machine learning researchers to move away from neural networks, and towards Gaussian processes.

### 18.2.7 Exercises

1. Draw sample prior functions from a GP with an Ornstein-Uhlenbeck (OU) kernel, $k_{\textrm{OU}}(x, x') = \exp\left(-\frac{1}{2\ell}||x - x'|\right)$. If you fix the lengthscale $\ell$ to be the same, how do these functions look different than sample functions from a GP with an RBF kernel?

2. How does changing the *amplitude* $a^2$ of the RBF kernel affect the distribution over functions?

3. Suppose we form $u(x) = f(x) + 2g(x)$, where $f(x) \sim \mathcal{GP}(m_1, k_1)$ and $g(x) \sim \mathcal{GP}(m_2, k_2)$. Is $u(x)$ a Gaussian process, and if so, what is its mean and covariance function?

4. Suppose we form $g(x) = a(x)f(x)$, where $f(x) \sim \mathcal{GP}(0, k)$ and $a(x) = x^2$. Is $g(x)$ a Gaussian process, and if so, what is its mean and covariance function? What is the effect of $a(x)$? What do sample functions drawn from $g(x)$ look like?

5. Suppose we form $u(x) = f(x)g(x)$, where $f(x) \sim \mathcal{GP}(m_1, k_1)$ and $g(x) \sim \mathcal{GP}(m_2, k_2)$. Is $u(x)$ a Gaussian process, and if so, what is its mean and covariance function?

Discussions[260].

# 18.3 Gaussian Process Inference

In this section, we will show how to perform posterior inference and make predictions using the GP priors we introduced in the last section. We will start with regression, where we can perform inference in *closed form*. This is a "GPs in a nutshell" section to quickly get up and running with Gaussian processes in practice. We'll start coding all the basic operations from scratch, and then introduce GPyTorch[261] , which will make working with state-of-the-art Gaussian processes and integration with deep neural networks much more convenient. We will consider these more advanced topics in depth in the next section. In that section, we will also consider settings where approximate inference is required — classification, point processes, or any non-Gaussian likelihoods.

## 18.3.1 Posterior Inference for Regression

An *observation* model relates the function we want to learn, $f(x)$, to our observations $y(x)$, both indexed by some input $x$. In classification, $x$ could be the pixels of an image, and $y$ could be the associated class label. In regression, $y$ typically represents a continuous output, such as a land surface temperature, a sea-level, a $CO_2$ concentration, etc.

In regression, we often assume the outputs are given by a latent noise-free function $f(x)$ plus i.i.d. Gaussian noise $\epsilon(x)$:

$$y(x) = f(x) + \epsilon(x), \tag{18.3.1}$$

with $\epsilon(x) \sim \mathcal{N}(0, \sigma^2)$. Let $\mathbf{y} = y(X) = (y(x_1), \ldots, y(x_n))^\top$ be a vector of our training observations, and $\mathbf{f} = (f(x_1), \ldots, f(x_n))^\top$ be a vector of the latent noise-free function values, queried at the training inputs $X = x_1, \ldots, x_n$.

We will assume $f(x) \sim \mathcal{GP}(m, k)$, which means that any collection of function values $\mathbf{f}$ has a joint multivariate Gaussian distribution, with mean vector $\mu_i = m(x_i)$ and covariance matrix $K_{ij} = k(x_i, x_j)$. The RBF kernel $k(x_i, x_j) = a^2 \exp\left(-\frac{1}{2\ell^2}||x_i - x_j||^2\right)$ would be a standard choice of covariance function. For notational simplicity, we will assume the mean function $m(x) = 0$; our derivations can easily be generalized later on.

Suppose we want to make predictions at a set of inputs

$$X_* = x_{*1}, x_{*2}, \ldots, x_{*m}. \tag{18.3.2}$$

Then we want to find $x^2$ and $p(\mathbf{f}_*|\mathbf{y}, X)$. In the regression setting, we can conveniently find this distribution by using Gaussian identities, after finding the joint distribution over $\mathbf{f}_* = f(X_*)$ and $\mathbf{y}$.

If we evaluate equation (18.3.1) at the training inputs $X$, we have $\mathbf{y} = \mathbf{f} +$ ffl. By the definition of a Gaussian process (see last section), $\mathbf{f} \sim \mathcal{N}(0, K(X, X))$ where $K(X, X)$ is an $n \times n$ matrix formed by evaluating our covariance function (aka *kernel*) at all possible pairs of inputs $x_i, x_j \in X$. ﬄ is simply a vector comprised of iid samples from $\mathcal{N}(0, \sigma^2)$ and thus has distribution $\mathcal{N}(0, \sigma^2 I)$. $\mathbf{y}$ is therefore a sum of two independent multivariate Gaussian variables, and thus has distribution $\mathcal{N}(0, K(X, X) + \sigma^2 I)$. One can also show that $\mathrm{cov}(\mathbf{f}_*, \mathbf{y}) = \mathrm{cov}(\mathbf{y}, \mathbf{f}_*)^\top = K(X_*, X)$ where $K(X_*, X)$ is an $m \times n$ matrix formed by evaluating the kernel at all pairs of test and training inputs.

$$\begin{bmatrix} \mathbf{y} \\ \mathbf{f}_* \end{bmatrix} \sim \mathcal{N}\left(0, \mathbf{A} = \begin{bmatrix} K(X,X)+\sigma^2 I & K(X,X_*) \\ K(X_*,X) & K(X_*,X_*) \end{bmatrix}\right) \tag{18.3.3}$$

We can then use standard Gaussian identities to find the conditional distribution from the joint distribution (see, e.g., Bishop Chapter 2), $\mathbf{f}_* | \mathbf{y}, X, X_* \sim \mathcal{N}(m_*, S_*)$, where $m_* = K(X_*, X)[K(X, X) + \sigma^2 I]^{-1}\mathbf{y}$, and $S = K(X_*, X_*) - K(X_*, X)[K(X, X) + \sigma^2 I]^{-1} K(X, X_*)$.

Typically, we do not need to make use of the full predictive covariance matrix $S$, and instead use the diagonal of $S$ for uncertainty about each prediction. Often for this reason we write the predictive distribution for a single test point $x_*$, rather than a collection of test points.

The kernel matrix has parameters $\theta$ that we also wish to estimate, such the amplitude $a$ and lengthscale $\ell$ of the RBF kernel above. For these purposes we use the *marginal likelihood*, $p(\mathbf{y}|\theta, X)$, which we already derived in working out the marginal distributions to find the joint distribution over $\mathbf{y}, \mathbf{f}_*$. As we will see, the marginal likelihood compartmentalizes into model fit and model complexity terms, and automatically encodes a notion of Occam's razor for learning hyperparameters. For a full discussion, see MacKay Ch. 28 (MacKay, 2003), and Rasmussen and Williams Ch. 5 (Rasmussen and Williams, 2006).

```
import math
import os
import gpytorch
import matplotlib.pyplot as plt
import numpy as np
import torch
from scipy import optimize
from scipy.spatial import distance_matrix
from d2l import torch as d2l

d2l.set_figsize()
```

## 18.3.2 Equations for Making Predictions and Learning Kernel Hyperparameters in GP Regression

We list here the equations you will use for learning hyperparameters and making predictions in Gaussian process regression. Again, we assume a vector of regression targets $\mathbf{y}$, indexed by inputs $X = \{x_1, \ldots, x_n\}$, and we wish to make a prediction at a test input $x_*$. We assume i.i.d. additive zero-mean Gaussian noise with variance $\sigma^2$. We use a Gaussian process prior $f(x) \sim \mathcal{GP}(m, k)$ for the latent noise-free function, with mean function $m$ and kernel function $k$. The kernel itself has parameters $\theta$ that we want to learn. For example, if we use an RBF kernel, $k(x_i, x_j) = a^2 \exp\left(-\frac{1}{2\ell^2}||x - x'||^2\right)$, we want to learn $\theta = \{a^2, \ell^2\}$. Let $K(X, X)$ represent an $n \times n$ matrix corresponding to evaluating the kernel for all possible pairs of $n$ training inputs. Let $K(x_*, X)$ represent a $1 \times n$ vector formed by evaluating $k(x_*, x_i)$, $i = 1, \ldots, n$. Let $\mu$ be a mean vector formed by evaluating the mean function $m(x)$ at every training points $x$.

Typically in working with Gaussian processes, we follow a two-step procedure. 1. Learn kernel hyperparameters $\hat{\theta}$ by maximizing the marginal likelihood with respect to these hyperparameters. 2. Use the predictive mean as a point predictor, and 2 times the predictive standard deviation to form a 95% credible set, conditioning on these learned hyperparameters $\hat{\theta}$.

The log marginal likelihood is simply a log Gaussian density, which has the form:

$$\log p(\mathbf{y}|\theta, X) = -\frac{1}{2}\mathbf{y}^\top [K_\theta(X, X) + \sigma^2 I]^{-1}\mathbf{y} - \frac{1}{2}\log|K_\theta(X, X)| + c \tag{18.3.4}$$

The predictive distribution has the form:

$$p(y_*|x_*, \mathbf{y}, \theta) = \mathcal{N}(a_*, v_*) \tag{18.3.5}$$

$$a_* = k_\theta(x_*, X)[K_\theta(X, X) + \sigma^2 I]^{-1}(\mathbf{y} - \mu) + \mu \tag{18.3.6}$$

$$v_* = k_\theta(x_*, x_*) - K_\theta(x_*, X)[K_\theta(X, X) + \sigma^2 I]^{-1} k_\theta(X, x_*) \tag{18.3.7}$$

## 18.3.3 Interpreting Equations for Learning and Predictions

There are some key points to note about the predictive distributions for Gaussian processes:

- Despite the flexibility of the model class, it is possible to do *exact* Bayesian inference for GP regression in *closed form*. Aside from learning the kernel hyperparameters, there is no *training*. We can write down exactly what equations we want to use to make predictions. Gaussian processes are relatively exceptional in this respect, and it has greatly contributed to their convenience, versatility, and continued popularity.
- The predictive mean $a_*$ is a linear combination of the training targets $\mathbf{y}$, weighted by the kernel $k_\theta(x_*, X)[K_\theta(X, X) + \sigma^2 I]^{-1}$. As we will see, the kernel (and its hyperparameters) thus plays a crucial role in the generalization properties of the model.
- The predictive mean explicitly depends on the target values $\mathbf{y}$ but the predictive variance does not. The predictive uncertainty instead grows as the test input $x_*$ moves away from the target locations $X$, as governed by the kernel function. However, uncertainty will implicitly depend on the values of the targets $\mathbf{y}$ through the kernel hyperparameters $\theta$, which are learned from the data.
- The marginal likelihood compartmentalizes into model fit and model complexity (log determinant) terms. The marginal likelihood tends to select for hyperparameters that provide the simplest fits that are still consistent with the data.
- The key computational bottlenecks come from solving a linear system and computing a log determinant over an $n \times n$ symmetric positive definite matrix $K(X, X)$ for $n$ training points. Naively, these operations each incur $O(n^3)$ computations, as well as $O(n^2)$ storage for each entry of the kernel (covariance) matrix, often starting with a Cholesky decomposition. Historically, these bottlenecks have limited GPs to problems with fewer than about 10,000 training points, and have given GPs a reputation for "being slow" that has been inaccurate now for almost a decade. In advanced topics, we will discuss how GPs can be scaled to problems with millions of points.

- For popular choices of kernel functions, $K(X, X)$ is often close to singular, which can cause numerical issues when performing Cholesky decompositions or other operations intended to solve linear systems. Fortunately, in regression we are often working with $K_\theta(X, X) + \sigma^2 I$, such that the noise variance $\sigma^2$ gets added to the diagonal of $K(X, X)$, significantly improving its conditioning. If the noise variance is small, or we are doing noise free regression, it is common practice to add a small amount of "jitter" to the diagonal, on the order of $10^{-6}$, to improve conditioning.

### 18.3.4 Worked Example from Scratch

Let's create some regression data, and then fit the data with a GP, implementing every step from scratch. We'll sample data from

$$y(x) = \sin(x) + \frac{1}{2}\sin(4x) + \epsilon, \tag{18.3.8}$$

with $\epsilon \sim \mathcal{N}(0, \sigma^2)$. The noise free function we wish to find is $f(x) = \sin(x) + \frac{1}{2}\sin(4x)$. We'll start by using a noise standard deviation $\sigma = 0.25$.

```
def data_maker1(x, sig):
    return np.sin(x) + 0.5 * np.sin(4 * x) + np.random.randn(x.shape[0]) * sig

sig = 0.25
train_x, test_x = np.linspace(0, 5, 50), np.linspace(0, 5, 500)
train_y, test_y = data_maker1(train_x, sig=sig), data_maker1(test_x, sig=0.)

d2l.plt.scatter(train_x, train_y)
d2l.plt.plot(test_x, test_y)
d2l.plt.xlabel("x", fontsize=20)
d2l.plt.ylabel("Observations y", fontsize=20)
d2l.plt.show()
```

Here we see the noisy observations as circles, and the noise-free function in blue that we wish to find.

Now, let's specify a GP prior over the latent noise-free function, $f(x) \sim \mathcal{GP}(m, k)$. We'll use a mean function $m(x) = 0$, and an RBF covariance function (kernel)

$$k(x_i, x_j) = a^2 \exp\left(-\frac{1}{2\ell^2}||x - x'||^2\right). \tag{18.3.9}$$

```
mean = np.zeros(test_x.shape[0])
cov = d2l.rbfkernel(test_x, test_x, ls=0.2)
```

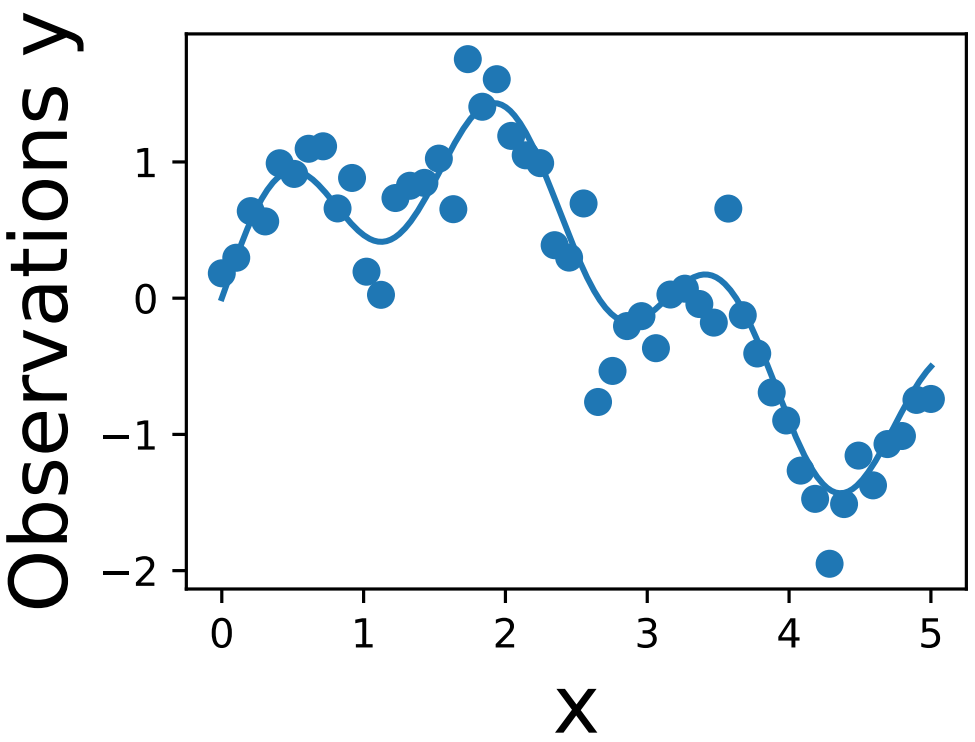

We have started with a length-scale of 0.2. Before we fit the data, it is important to consider whether we have specified a reasonable prior. Let's visualize some sample functions from this prior, as well as the 95% credible set (we believe there's a 95% chance that the true function is within this region).

```
prior_samples = np.random.multivariate_normal(mean=mean, cov=cov, size=5)
d2l.plt.plot(test_x, prior_samples.T, color='black', alpha=0.5)
d2l.plt.plot(test_x, mean, linewidth=2.)
d2l.plt.fill_between(test_x, mean - 2 * np.diag(cov), mean + 2 * np.diag(cov),
                     alpha=0.25)
d2l.plt.show()
```

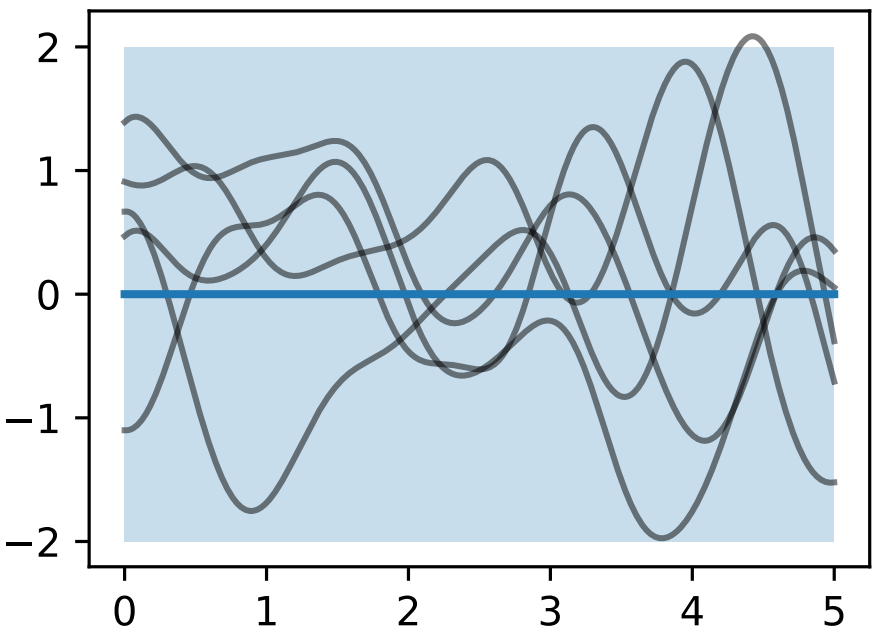

Do these samples look reasonable? Are the high-level properties of the functions aligned with the type of data we are trying to model?

Now let's form the mean and variance of the posterior predictive distribution at any arbitrary test point $x_*$.

$$\bar{f}_* = K(x, x_*)^T (K(x, x) + \sigma^2 I)^{-1} y \tag{18.3.10}$$

$$V(f_*) = K(x_*, x_*) - K(x, x_*)^T (K(x, x) + \sigma^2 I)^{-1} K(x, x_*) \tag{18.3.11}$$

Before we make predictions, we should learn our kernel hyperparameters $\theta$ and noise variance $\sigma^2$. Let's initialize our length-scale at 0.75, as our prior functions looked too quickly varying compared to the data we are fitting. We'll also guess a noise standard deviation $\sigma$ of 0.75.

In order to learn these parameters, we will maximize the marginal likelihood with respect to these parameters.

$$\log p(y|X) = \log \int p(y|f, X)p(f|X)df \tag{18.3.12}$$

$$\log p(y|X) = -\frac{1}{2}y^T(K(x,x) + \sigma^2 I)^{-1}y - \frac{1}{2}\log|K(x,x) + \sigma^2 I| - \frac{n}{2}\log 2\pi \tag{18.3.13}$$

Perhaps our prior functions were too quickly varying. Let's guess a length-scale of 0.4. We'll also guess a noise standard deviation of 0.75. These are simply hyperparameter initializations — we will learn these parameters from the marginal likelihood.

```
ell_est = 0.4
post_sig_est = 0.5

def neg_MLL(pars):
    K = d2l.rbfkernel(train_x, train_x, ls=pars[0])
    kernel_term = -0.5 * train_y @ \
        np.linalg.inv(K + pars[1] ** 2 * np.eye(train_x.shape[0])) @ train_y
    logdet = -0.5 * np.log(np.linalg.det(K + pars[1] ** 2 * \
                                         np.eye(train_x.shape[0])))
    const = -train_x.shape[0] / 2. * np.log(2 * np.pi)

    return -(kernel_term + logdet + const)

learned_hypers = optimize.minimize(neg_MLL, x0=np.array([ell_est,post_sig_
↪est]),
                                   bounds=((0.01, 10.), (0.01, 10.)))
ell = learned_hypers.x[0]
post_sig_est = learned_hypers.x[1]
```

In this instance, we learn a length-scale of 0.299, and a noise standard deviation of 0.24. Note that the learned noise is extremely close to the true noise, which helps indicate that our GP is a very well-specified to this problem.

In general, it is crucial to put careful thought into selecting the kernel and initializing the hyperparameters. While marginal likelihood optimization can be relatively robust to initialization, it is not immune to poor initializations. Try running the above script with a variety of initializations and see what results you find.

Now, let's make predictions with these learned hypers.

```
K_x_xstar = d2l.rbfkernel(train_x, test_x, ls=ell)
K_x_x = d2l.rbfkernel(train_x, train_x, ls=ell)
K_xstar_xstar = d2l.rbfkernel(test_x, test_x, ls=ell)

post_mean = K_x_xstar.T @ np.linalg.inv((K_x_x + \
                post_sig_est ** 2 * np.eye(train_x.shape[0]))) @ train_y
```

(continued from previous page)

```
post_cov = K_xstar_xstar - K_x_xstar.T @ np.linalg.inv((K_x_x + \
                post_sig_est ** 2 * np.eye(train_x.shape[0]))) @ K_x_xstar

lw_bd = post_mean - 2 * np.sqrt(np.diag(post_cov))
up_bd = post_mean + 2 * np.sqrt(np.diag(post_cov))

d2l.plt.scatter(train_x, train_y)
d2l.plt.plot(test_x, test_y, linewidth=2.)
d2l.plt.plot(test_x, post_mean, linewidth=2.)
d2l.plt.fill_between(test_x, lw_bd, up_bd, alpha=0.25)
d2l.plt.legend(['Observed Data', 'True Function', 'Predictive Mean', '95% Set␣
↪on True Func'])
d2l.plt.show()
```

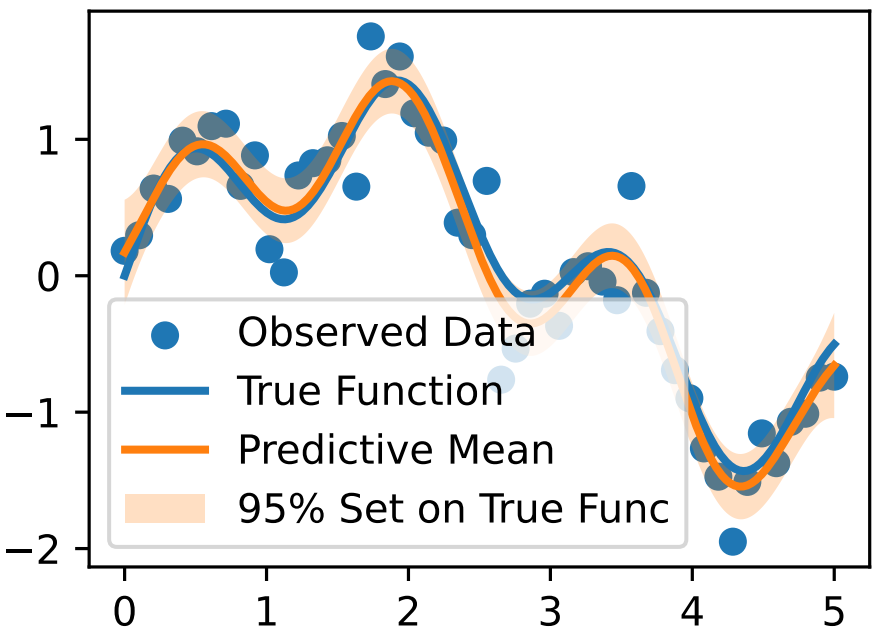

We see the posterior mean in orange almost perfectly matches the true noise free function! Note that the 95% credible set we are showing is for the latent *noise free* (true) function, and not the data points. We see that this credible set entirely contains the true function, and does not seem overly wide or narrow. We would not want nor expect it to contain the data points. If we wish to have a credible set for the observations, we should compute

```
lw_bd_observed = post_mean - 2 * np.sqrt(np.diag(post_cov) + post_sig_est ** 2)
up_bd_observed = post_mean + 2 * np.sqrt(np.diag(post_cov) + post_sig_est ** 2)
```

There are two sources of uncertainty, *epistemic* uncertainty, representing *reducible* uncertainty, and *aleatoric* or *irreducible* uncertainty. The *epistemic* uncertainty here represents uncertainty about the true values of the noise free function. This uncertainty should grow as we move away from the data points, as away from the data there are a greater variety of function values consistent with our data. As we observe more and more data, our beliefs about the true function become more confident, and the epistemic uncertainty disappears. The *aleatoric* uncertainty in this instance is the observation noise, since the data are given to us with this noise, and it cannot be reduced.

The *epistemic* uncertainty in the data is captured by variance of the latent noise free function np.diag(post_cov). The *aleatoric* uncertainty is captured by the noise variance post_sig_est**2.

Unfortunately, people are often careless about how they represent uncertainty, with many papers showing error bars that are completely undefined, no clear sense of whether we are visualizing epistemic or aleatoric uncertainty or both, and confusing noise variances with noise standard deviations, standard deviations with standard errors, confidence intervals with credible sets, and so on. Without being precise about what the uncertainty represents, it is essentially meaningless.

In the spirit of playing close attention to what our uncertainty represents, it is crucial to note that we are taking *two times* the *square root* of our variance estimate for the noise free function. Since our predictive distribution is Gaussian, this quantity enables us to form a 95% credible set, representing our beliefs about the interval which is 95% likely to contain the ground truth function. The noise *variance* is living on a completely different scale, and is much less interpretable.

Finally, let's take a look at 20 posterior samples. These samples tell us what types of functions we believe might fit our data, a posteriori.

```
post_samples = np.random.multivariate_normal(post_mean, post_cov, size=20)
d2l.plt.scatter(train_x, train_y)
d2l.plt.plot(test_x, test_y, linewidth=2.)
d2l.plt.plot(test_x, post_mean, linewidth=2.)
d2l.plt.plot(test_x, post_samples.T, color='gray', alpha=0.25)
d2l.plt.fill_between(test_x, lw_bd, up_bd, alpha=0.25)
plt.legend(['Observed Data', 'True Function', 'Predictive Mean', 'Posterior␣
↪Samples'])
d2l.plt.show()
```

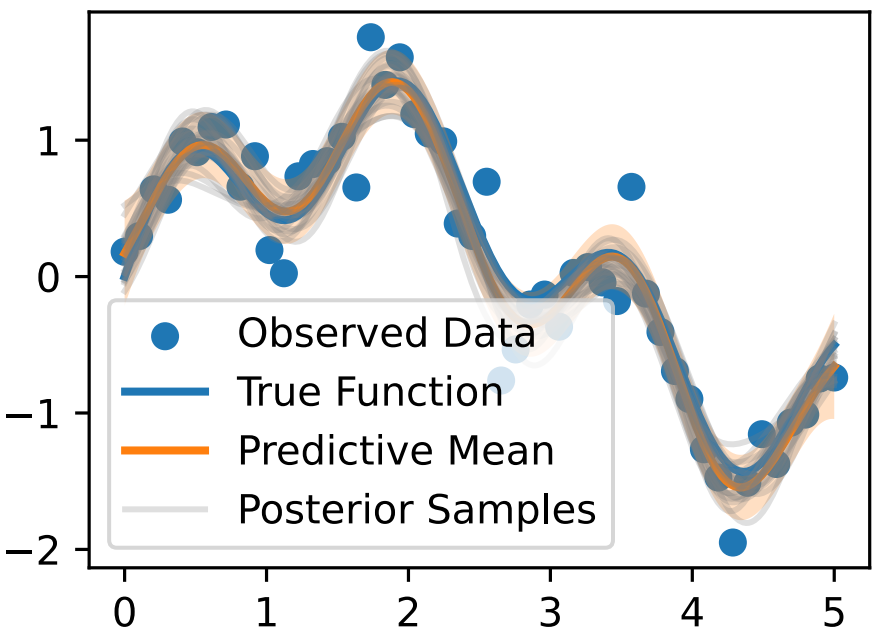

In basic regression applications, it is most common to use the posterior predictive mean and standard deviation as a point predictor and metric for uncertainty, respectively. In more advanced applications, such as Bayesian optimization with Monte Carlo acquisition functions, or Gaussian processes for model-based RL, it often necessary to take posterior samples. However, even if not strictly required in the basic applications, these samples give us more intuition about the fit we have for the data, and are often useful to include in visualizations.

### 18.3.5 Making Life Easy with GPyTorch

As we have seen, it is actually pretty easy to implement basic Gaussian process regression entirely from scratch. However, as soon as we want to explore a variety of kernel choices, consider approximate inference (which is needed even for classification), combine GPs with neural networks, or even have a dataset larger than about 10,000 points, then an implementation from scratch becomes unwieldy and cumbersome. Some of the most effective methods for scalable GP inference, such as SKI (also known as KISS-GP), can require hundreds of lines of code implementing advanced numerical linear algebra routines.

In these cases, the *GPyTorch* library will make our lives a lot easier. We'll be discussing GPyTorch more in future notebooks on Gaussian process numerics, and advanced methods. The GPyTorch library contains many examples[262]. To get a feel for the package, we will walk through the simple regression example[263], showing how it can be adapted to reproduce our above results using GPyTorch. This may seem like a lot of code to simply reproduce the basic regression above, and in a sense, it is. But we can immediately use a variety of kernels, scalable inference techniques, and approximate inference, by only changing a few lines of code from below, instead of writing potentially thousands of lines of new code.

```
# First let's convert our data into tensors for use with PyTorch
train_x = torch.tensor(train_x)
train_y = torch.tensor(train_y)
test_y = torch.tensor(test_y)

# We are using exact GP inference with a zero mean and RBF kernel
class ExactGPModel(gpytorch.models.ExactGP):
    def __init__(self, train_x, train_y, likelihood):
        super(ExactGPModel, self).__init__(train_x, train_y, likelihood)
        self.mean_module = gpytorch.means.ZeroMean()
        self.covar_module = gpytorch.kernels.ScaleKernel(
            gpytorch.kernels.RBFKernel())

    def forward(self, x):
        mean_x = self.mean_module(x)
        covar_x = self.covar_module(x)
        return gpytorch.distributions.MultivariateNormal(mean_x, covar_x)
```

This code block puts the data in the right format for GPyTorch, and specifies that we are using exact inference, as well the mean function (zero) and kernel function (RBF) that we want to use. We can use any other kernel very easily, by calling, for instance, gpytorch.kernels.matern_kernel(), or gpyotrch.kernels.spectral_mixture_kernel(). So far, we have only discussed exact inference, where it is possible to infer a predictive distribution without making any approximations. For Gaussian processes, we can only perform exact inference when we have a Gaussian likelihood; more specifically, when we assume that our observations are generated as a noise-free function represented by a Gaussian process, plus Gaussian noise. In future notebooks, we will consider other settings, such as classification, where we cannot make these assumptions.

```
# Initialize Gaussian likelihood
likelihood = gpytorch.likelihoods.GaussianLikelihood()
```

```
model = ExactGPModel(train_x, train_y, likelihood)
training_iter = 50
# Find optimal model hyperparameters
model.train()
likelihood.train()
# Use the adam optimizer, includes GaussianLikelihood parameters
optimizer = torch.optim.Adam(model.parameters(), lr=0.1)
# Set our loss as the negative log GP marginal likelihood
mll = gpytorch.mlls.ExactMarginalLogLikelihood(likelihood, model)
```

Here, we explicitly specify the likelihood we want to use (Gaussian), the objective we will use for training kernel hyperparameters (here, the marginal likelihood), and the procedure we we want to use for optimizing that objective (in this case, Adam). We note that while we are using Adam, which is a "stochastic" optimizer, in this case, it is full-batch Adam. Because the marginal likelihood does not factorize over data instances, we cannot use an optimizer over "mini-batches" of data and be guaranteed convergence. Other optimizers, such as L-BFGS, are also supported by GPyTorch. Unlike in standard deep learning, doing a good job of optimizing the marginal likelihood corresponds strongly with good generalization, which often inclines us towards powerful optimizers like L-BFGS, assuming they are not prohibitively expensive.

```
for i in range(training_iter):
    # Zero gradients from previous iteration
    optimizer.zero_grad()
    # Output from model
    output = model(train_x)
    # Calc loss and backprop gradients
    loss = -mll(output, train_y)
    loss.backward()
    if i % 10 == 0:
        print(f'Iter {i+1:d}/{training_iter:d} - Loss: {loss.item():.3f} '
              f'squared lengthscale: '
              f'{model.covar_module.base_kernel.lengthscale.item():.3f} '
              f'noise variance: {model.likelihood.noise.item():.3f}')
    optimizer.step()
```

```
Iter 1/50 - Loss: 1.013 squared lengthscale: 0.693 noise variance: 0.693
Iter 11/50 - Loss: 0.739 squared lengthscale: 0.508 noise variance: 0.312
Iter 21/50 - Loss: 0.522 squared lengthscale: 0.515 noise variance: 0.129
Iter 31/50 - Loss: 0.474 squared lengthscale: 0.511 noise variance: 0.064
Iter 41/50 - Loss: 0.481 squared lengthscale: 0.505 noise variance: 0.058
```

Here we actually run the optimization procedure, outputting the values of the loss every 10 iterations.

```
# Get into evaluation (predictive posterior) mode
test_x = torch.tensor(test_x)
model.eval()
```

```
likelihood.eval()
observed_pred = likelihood(model(test_x))
```

The above codeblock enables us to make predictions on our test inputs.

```
with torch.no_grad():
    # Initialize plot
    f, ax = d2l.plt.subplots(1, 1, figsize=(4, 3))
    # Get upper and lower bounds for 95\% credible set (in this case, in
    # observation space)
    lower, upper = observed_pred.confidence_region()
    ax.scatter(train_x.numpy(), train_y.numpy())
    ax.plot(test_x.numpy(), test_y.numpy(), linewidth=2.)
    ax.plot(test_x.numpy(), observed_pred.mean.numpy(), linewidth=2.)
    ax.fill_between(test_x.numpy(), lower.numpy(), upper.numpy(), alpha=0.25)
    ax.set_ylim([-1.5, 1.5])
    ax.legend(['True Function', 'Predictive Mean', 'Observed Data',
               '95% Credible Set'])
```

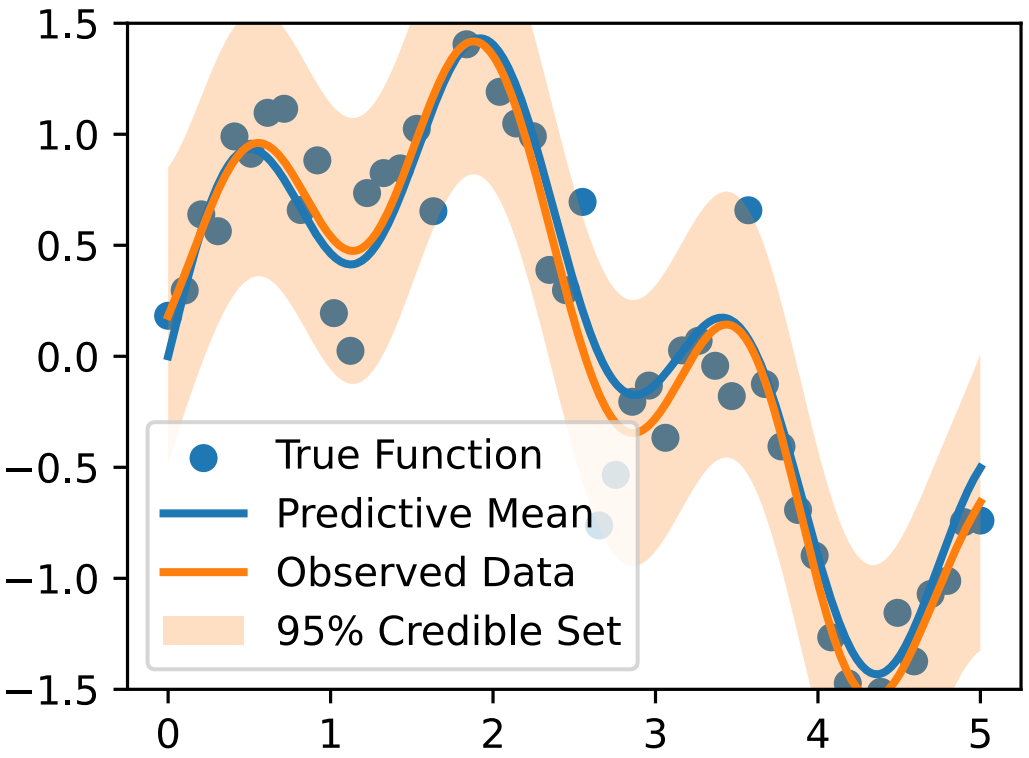

Finally, we plot the fit.

We see the fits are virtually identical. A few things to note: GPyTorch is working with *squared* length-scales and observation noise. For example, our learned noise standard deviation in the for scratch code is about 0.283. The noise variance found by GPyTorch is $0.81 \approx 0.283^2$. In the GPyTorch plot, we also show the credible set in the *observation space* rather than the latent function space, to demonstrate that they indeed cover the observed datapoints.

## 18.3.6 Summary

We can combine a Gaussian process prior with data to form a posterior, which we use to make predictions. We can also form a marginal likelihood, which is useful for automatic learning of kernel hyperparameters, which control properties such as the rate of variation of the Gaussian process. The mechanics of forming the posterior and learning kernel hyperparameters for regression are simple, involving about a dozen lines of code. This notebook is a good reference for any reader wanting to quickly get “up and running” with Gaussian processes. We also introduced the GPyTorch library. Although the GPyTorch code for basic regression is relatively long, it can be trivially modified for other kernel functions, or more advanced functionality we will discuss in future notebooks, such as scalable inference, or non-Gaussian likelihoods for classification.

## 18.3.7 Exercises

1. We have emphasized the importance of *learning* kernel hyperparameters, and the effect of hyperparameters and kernels on the generalization properties of Gaussian processes. Try skipping the step where we learn hypers, and instead guess a variety of length-scales and noise variances, and check their effect on predictions. What happens when you use a large length-scale? A small length-scale? A large noise variance? A small noise variance?

2. We have said that the marginal likelihood is not a convex objective, but that hyperparameters like length-scale and noise variance can be reliably estimated in GP regression. This is generally true — in fact, the marginal likelihood is *much* better at learning length-scale hyperparameters than conventional approaches in spatial statistics, which involve fitting empirical autocorrelation functions (“covariograms”). Arguably, the biggest contribution from machine learning to Gaussian process research, at least before recent work on scalable inference, was the introduction of the marginal lkelihood for hyperparameter learning.

*However*, different pairings of even these parameters provide interpretably different plausible explanations for many datasets, leading to local optima in our objective. If we use a large length-scale, then we assume the true underlying function is slowly varying. If the observed data *are* varying significantly, then the only we can plausibly have a large length-scale is with a large noise-variance. If we use a small length-scale, on the other hand, our fit will be very sensitive to the variations in the data, leaving little room to explain variations with noise (aleatoric uncertainty).

Try seeing if you can find these local optima: initialize with very large length-scale with large noise, and small length-scales with small noise. Do you converge to different solutions?

3. We have said that a fundamental advantage of Bayesian methods is in naturally representing *epistemic* uncertainty. In the above example, we cannot fully see the effects of epistemic uncertainty. Try instead to predict with `test_x = np.linspace(0, 10, 1000)`. What happens to the 95% credible set as your predictions move beyond the data? Does it cover the true function in that interval? What happens if you only visualize aleatoric uncertainty in that region?

4. Try running the above example, but instead with 10,000, 20,000 and 40,000 training points, and measure the runtimes. How does the training time scale? Alternatively, how do the runtimes scale with the number of test points? Is it different for the predictive mean and the predictive variance? Answer this question both by theoretically working out the training and testing time complexities, and by running the code above with a different number of points.

5. Try running the GPyTorch example with different covariance functions, such as the Matern kernel. How do the results change? How about the spectral mixture kernel, found in the GPyTorch library? Are some easier to train the marginal likelihood than others? Are some more valuable for long-range versus short-range predictions?

6. In our GPyTorch example, we plotted the predictive distribution including observation noise, while in our "from scratch" example, we only included epistemic uncertainty. Re-do the GPyTorch example, but this time only plotting epistemic uncertainty, and compare to the from-scratch results. Do the predictive distributions now look the same? (They should.)

Discussions[264].

# 19 Hyperparameter Optimization

**Aaron Klein** (*Amazon*), **Matthias Seeger** (*Amazon*), and **Cedric Archambeau** (*Amazon*)

The performance of every machine learning model depends on its hyperparameters. They control the learning algorithm or the structure of the underlying statistical model. However, there is no general way to choose hyperparameters in practice. Instead, hyperparameters are often set in a trial-and-error manner or sometimes left to their default values by practitioners, leading to suboptimal generalization.

Hyperparameter optimization provides a systematic approach to this problem, by casting it as an optimization problem: a good set of hyperparameters should (at least) minimize a validation error. Compared to most other optimization problems arising in machine learning, hyperparameter optimization is a nested one, where each iteration requires training and validating a machine learning model.

In this chapter, we will first introduce the basics of hyperparameter optimization. We will also present some recent advancements that improve the overall efficiency of hyperparameter optimization by exploiting cheap-to-evaluate proxies of the original objective function. At the end of this chapter, you should be able to apply state-of-the-art hyperparameter optimization techniques to optimize the hyperparameter of your own machine learning algorithm.

## 19.1 What Is Hyperparameter Optimization?

As we have seen in the previous chapters, deep neural networks come with a large number of parameters or weights that are learned during training. On top of these, every neural network has additional *hyperparameters* that need to be configured by the user. For example, to ensure that stochastic gradient descent converges to a local optimum of the training loss (see Chapter 12), we have to adjust the learning rate and batch size. To avoid overfitting on training datasets, we might have to set regularization parameters, such as weight decay (see Section 3.7) or dropout (see Section 5.6). We can define the capacity and inductive bias of the model by setting the number of layers and number of units or filters per layer (i.e., the effective number of weights).

Unfortunately, we cannot simply adjust these hyperparameters by minimizing the training loss, because this would lead to overfitting on the training data. For example, setting regularization parameters, such as dropout or weight decay to zero leads to a small training loss, but might hurt the generalization performance.

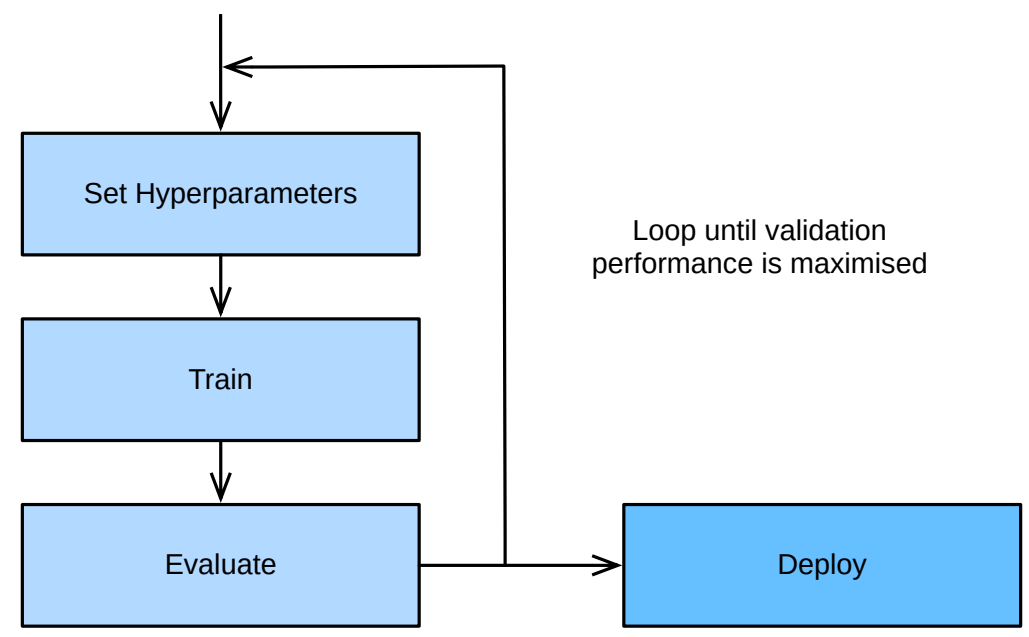

Fig. 19.1.1 Typical workflow in machine learning that consists of training the model multiple times with different hyperparameters.

Without a different form of automation, hyperparameters have to be set manually in a trial-and-error fashion, in what amounts to a time-consuming and difficult part of machine learning workflows. For example, consider training a ResNet (see Section 8.6) on CIFAR-10, which requires more than 2 hours on an Amazon Elastic Cloud Compute (EC2) `g4dn.xlarge` instance. Even just trying ten hyperparameter configurations in sequence, this would already take us roughly one day. To make matters worse, hyperparameters are usually not directly transferable across architectures and datasets (Bardenet *et al.*, 2013, Feurer *et al.*, 2022, Wistuba *et al.*, 2018), and need to be re-optimized for every new task. Also, for most hyperparameters, there are no rule-of-thumbs, and expert knowledge is required to find sensible values.

*Hyperparameter optimization (HPO)* algorithms are designed to tackle this problem in a principled and automated fashion (Feurer and Hutter, 2018), by framing it as a global optimization problem. The default objective is the error on a hold-out validation dataset, but could in principle be any other business metric. It can be combined with or constrained by secondary objectives, such as training time, inference time, or model complexity.

Recently, hyperparameter optimization has been extended to *neural architecture search (NAS)* (Elsken *et al.*, 2018, Wistuba *et al.*, 2019), where the goal is to find entirely new neural network architectures. Compared to classical HPO, NAS is even more expensive in terms of computation and requires additional efforts to remain feasible in practice. Both, HPO and NAS can be considered as sub-fields of AutoML (Hutter *et al.*, 2019), which aims to automate the entire ML pipeline.

In this section we will introduce HPO and show how we can automatically find the best hyperparameters of the logistic regression example introduced in Section 4.5.

### 19.1.1 The Optimization Problem

We will start with a simple toy problem: searching for the learning rate of the multi-class logistic regression model `SoftmaxRegression` from Section 4.5 to minimize the validation error on the Fashion MNIST dataset. While other hyperparameters like batch size or number of epochs are also worth tuning, we focus on learning rate alone for simplicity.

```
import numpy as np
import torch
from scipy import stats
from torch import nn
from d2l import torch as d2l
```

Before we can run HPO, we first need to define two ingredients: the objective function and the configuration space.

### The Objective Function

The performance of a learning algorithm can be seen as a function $f : \mathcal{X} \rightarrow \mathbb{R}$ that maps from the hyperparameter space $\mathbf{x} \in \mathcal{X}$ to the validation loss. For every evaluation of $f(\mathbf{x})$, we have to train and validate our machine learning model, which can be time and compute intensive in the case of deep neural networks trained on large datasets. Given our criterion $f(\mathbf{x})$ our goal is to find $\mathbf{x}_{\star} \in \mathrm{argmin}_{\mathbf{x} \in \mathcal{X}} f(\mathbf{x})$.

There is no simple way to compute gradients of $f$ with respect to $\mathbf{x}$, because it would require to propagate the gradient through the entire training process. While there is recent work (Franceschi *et al.*, 2017, Maclaurin *et al.*, 2015) to drive HPO by approximate "hypergradients", none of the existing approaches are competitive with the state-of-the-art yet, and we will not discuss them here. Furthermore, the computational burden of evaluating $f$ requires HPO algorithms to approach the global optimum with as few samples as possible.

The training of neural networks is stochastic (e.g., weights are randomly initialized, mini-batches are randomly sampled), so that our observations will be noisy: $y \sim f(\mathbf{x}) + \epsilon$, where we usually assume that the $\epsilon \sim N(0, \sigma)$ observation noise is Gaussian distributed.

Faced with all these challenges, we usually try to identify a small set of well performing hyperparameter configurations quickly, instead of hitting the global optima exactly. However, due to large computational demands of most neural networks models, even this can take days or weeks of compute. We will explore in Section 19.4 how we can speed-up the optimization process by either distributing the search or using cheaper-to-evaluate approximations of the objective function.

We begin with a method for computing the validation error of a model.

```
class HPOTrainer(d2l.Trainer):  #@save
    def validation_error(self):
        self.model.eval()
        accuracy = 0
```

```
        val_batch_idx = 0
        for batch in self.val_dataloader:
            with torch.no_grad():
                x, y = self.prepare_batch(batch)
                y_hat = self.model(x)
                accuracy += self.model.accuracy(y_hat, y)
            val_batch_idx += 1
        return 1 -  accuracy / val_batch_idx
```

We optimize validation error with respect to the hyperparameter configuration `config`, consisting of the `learning_rate`. For each evaluation, we train our model for `max_epochs` epochs, then compute and return its validation error:

```
def hpo_objective_softmax_classification(config, max_epochs=8):
    learning_rate = config["learning_rate"]
    trainer = d2l.HPOTrainer(max_epochs=max_epochs)
    data = d2l.FashionMNIST(batch_size=16)
    model = d2l.SoftmaxRegression(num_outputs=10, lr=learning_rate)
    trainer.fit(model=model, data=data)
    return trainer.validation_error().detach().numpy()
```

### The Configuration Space

Along with the objective function $f(\mathbf{x})$, we also need to define the feasible set $\mathbf{x} \in \mathcal{X}$ to optimize over, known as *configuration space* or *search space*. For our logistic regression example, we will use:

```
config_space = {"learning_rate": stats.loguniform(1e-4, 1)}
```

Here we use the use the `loguniform` object from SciPy, which represents a uniform distribution between -4 and -1 in the logarithmic space. This object allows us to sample random variables from this distribution.

Each hyperparameter has a data type, such as `float` for `learning_rate`, as well as a closed bounded range (i.e., lower and upper bounds). We usually assign a prior distribution (e.g, uniform or log-uniform) to each hyperparameter to sample from. Some positive parameters, such as `learning_rate`, are best represented on a logarithmic scale as optimal values can differ by several orders of magnitude, while others, such as momentum, come with linear scale.

Below we show a simple example of a configuration space consisting of typical hyperparameters of a multi-layer perceptron including their type and standard ranges.

Table 19.1.1: Example configuration space of multi-layer perceptron

| Name | Type | Hyperparameter Ranges | log-scale |
| --- | --- | --- | --- |
| learning rate | float | $[10^{-6}, 10^{-1}]$ | yes |
| batch size | integer | $[8, 256]$ | yes |
| momentum | float | $[0, 0.99]$ | no |
| activation function | categorical | $\{\tanh, \text{relu}\}$ | • |
| number of units | integer | $[32, 1024]$ | yes |
| number of layers | integer | $[1, 6]$ | no |

In general, the structure of the configuration space $\mathcal{X}$ can be complex and it can be quite different from $\mathbb{R}^d$. In practice, some hyperparameters may depend on the value of others. For example, assume we try to tune the number of layers for a multi-layer perceptron, and for each layer the number of units. The number of units of the $l$-th layer is relevant only if the network has at least $l + 1$ layers. These advanced HPO problems are beyond the scope of this chapter. We refer the interested reader to (Baptista and Poloczek, 2018, Hutter *et al.*, 2011, Jenatton *et al.*, 2017).

The configuration space plays an important role for hyperparameter optimization, since no algorithms can find something that is not included in the configuration space. On the other hand, if the ranges are too large, the computation budget to find well performing configurations might become infeasible.

## 19.1.2 Random Search

*Random search* is the first hyperparameter optimization algorithm we will consider. The main idea of random search is to independently sample from the configuration space until a predefined budget (e.g maximum number of iterations) is exhausted, and to return the best observed configuration. All evaluations can be executed independently in parallel (see Section 19.3), but here we use a sequential loop for simplicity.

```
errors, values = [], []
num_iterations = 5

for i in range(num_iterations):
    learning_rate = config_space["learning_rate"].rvs()
    print(f"Trial {i}: learning_rate = {learning_rate}")
    y = hpo_objective_softmax_classification({"learning_rate": learning_rate})
    print(f"    validation_error = {y}")
    values.append(learning_rate)
    errors.append(y)
```

```
    validation_error = 0.21079999208450317
```

The best learning rate is then simply the one with the lowest validation error.

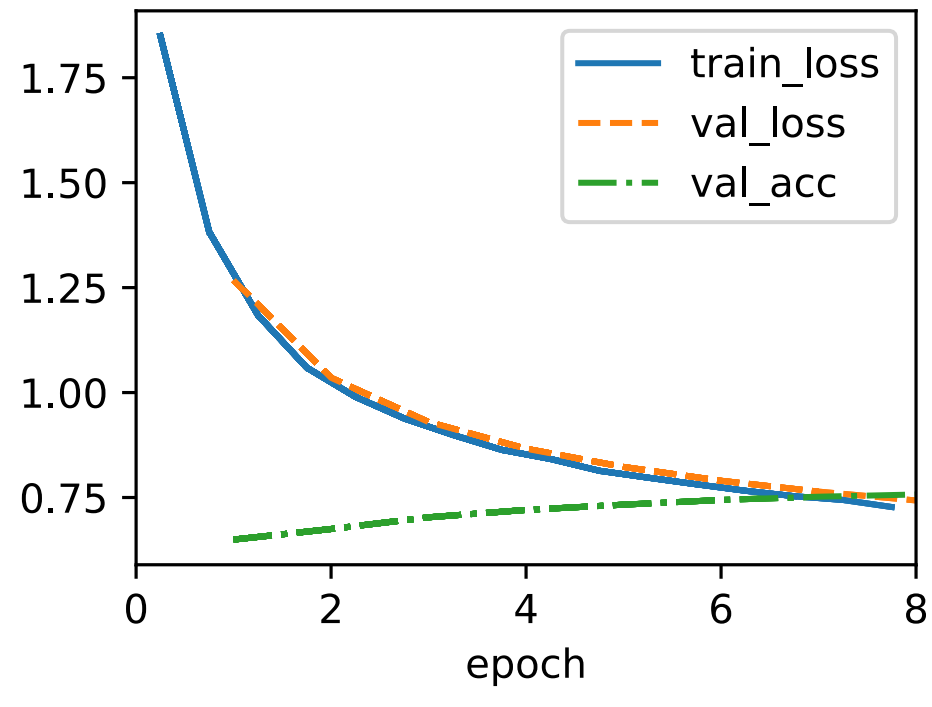

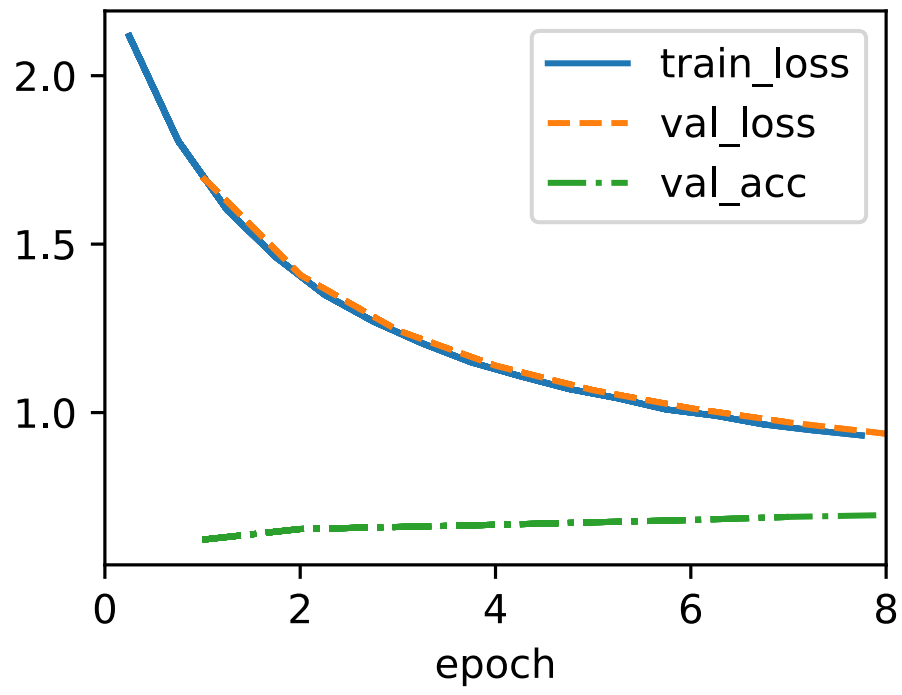

```
best_idx = np.argmin(errors)
print(f"optimal learning rate = {values[best_idx]}")
```

```
optimal learning rate = 0.05458176607692195
```

Due to its simplicity and generality, random search is one of the most frequently used HPO algorithms. It does not require any sophisticated implementation and can be applied to any configuration space as long as we can define some probability distribution for each hyperparameter.

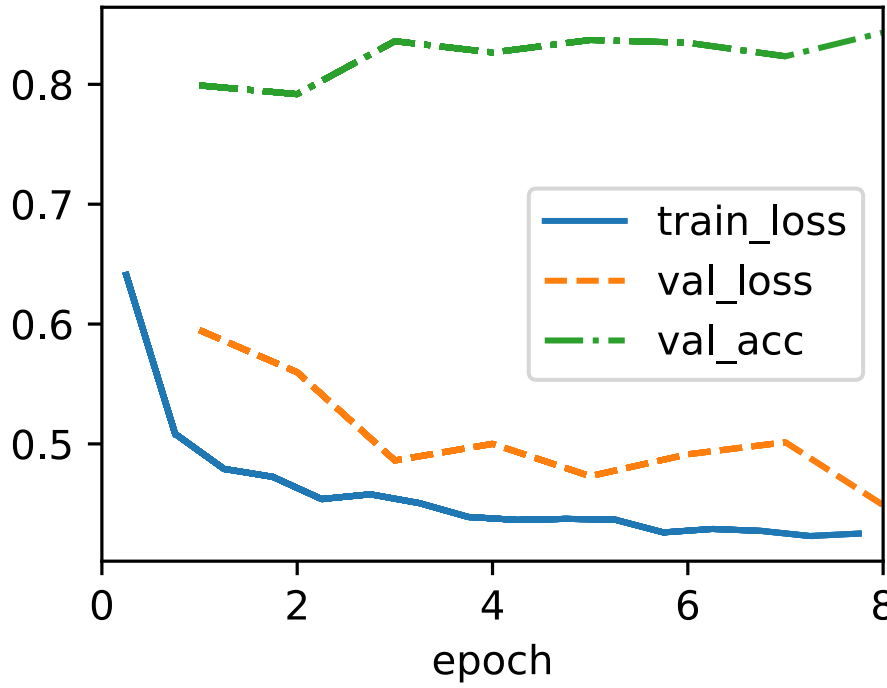

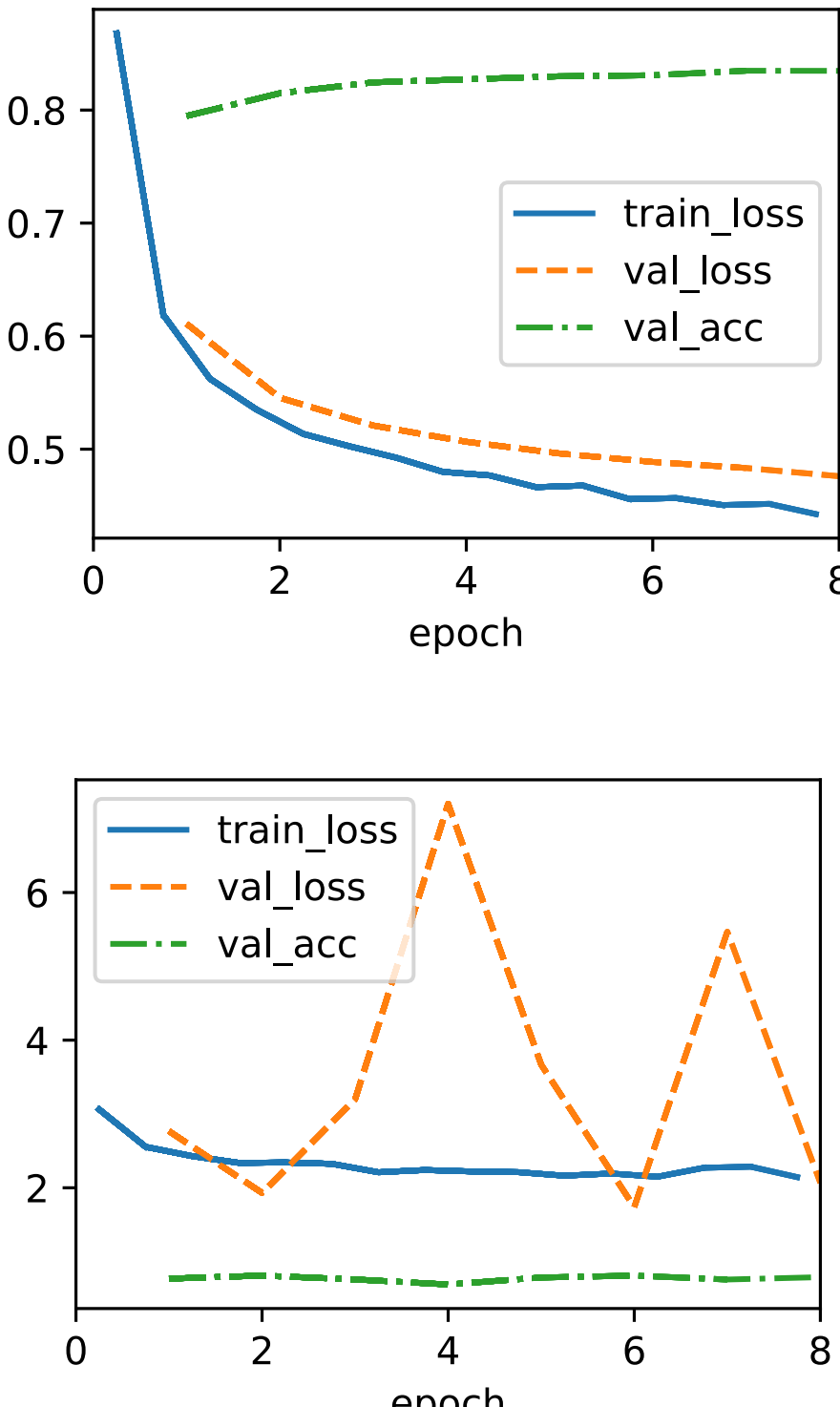

Unfortunately random search also comes with a few shortcomings. First, it does not adapt the sampling distribution based on the previous observations it collected so far. Hence, it is equally likely to sample a poorly performing configuration than a better performing configuration. Second, the same amount of resources are spent for all configurations, even though some may show poor initial performance and are less likely to outperform previously seen configurations.

In the next sections we will look at more sample efficient hyperparameter optimization algorithms that overcome the shortcomings of random search by using a model to guide the search. We will also look at algorithms that automatically stop the evaluation process of poorly performing configurations to speed up the optimization process.

### 19.1.3 Summary

In this section we introduced hyperparameter optimization (HPO) and how we can phrase it as a global optimization by defining a configuration space and an objective function. We also implemented our first HPO algorithm, random search, and applied it on a simple softmax classification problem.

While random search is very simple, it is the better alternative to grid search, which simply evaluates a fixed set of hyperparameters. Random search somewhat mitigates the curse of dimensionality (Bellman, 1966), and can be far more efficient than grid search if the criterion most strongly depends on a small subset of the hyperparameters.

### 19.1.4 Exercises

1. In this chapter, we optimize the validation error of a model after training on a disjoint training set. For simplicity, our code uses `Trainer.val_dataloader`, which maps to a loader around `FashionMNIST.val`.
    1. Convince yourself (by looking at the code) that this means we use the original FashionMNIST training set (60000 examples) for training, and the original *test set* (10000 examples) for validation.
    2. Why could this practice be problematic? Hint: Re-read Section 3.6, especially about *model selection*.
    3. What should we have done instead?
2. We stated above that hyperparameter optimization by gradient descent is very hard to do. Consider a small problem, such as training a two-layer perceptron on the FashionMNIST dataset (Section 5.2) with a batch size of 256. We would like to tune the learning rate of SGD in order to minimize a validation metric after one epoch of training.
    1. Why cannot we use validation *error* for this purpose? What metric on the validation set would you use?
    2. Sketch (roughly) the computational graph of the validation metric after training for one epoch. You may assume that initial weights and hyperparameters (such as learning rate) are input nodes to this graph. Hint: Re-read about computational graphs in Section 5.3.
    3. Give a rough estimate of the number of floating point values you need to store during a forward pass on this graph. Hint: FashionMNIST has 60000 cases. Assume the required memory is dominated by the activations after each layer, and look up the layer widths in Section 5.2.
    4. Apart from the sheer amount of compute and storage required, what other issues would gradient-based hyperparameter optimization run into? Hint: Re-read about vanishing and exploding gradients in Section 5.4.
    5. *Advanced*: Read (Maclaurin *et al.*, 2015) for an elegant (yet still somewhat unpractical) approach to gradient-based HPO.
3. Grid search is another HPO baseline, where we define an equi-spaced grid for each hyperparameter, then iterate over the (combinatorial) Cartesian product in order to suggest configurations.

1. We stated above that random search can be much more efficient than grid search for HPO on a sizable number of hyperparameters, if the criterion most strongly depends on a small subset of the hyperparameters. Why is this? Hint: Read (Bergstra *et al.*, 2011).

Discussions[265].

## 19.2 Hyperparameter Optimization API

Before we dive into the methodology, we will first discuss a basic code structure that allows us to efficiently implement various HPO algorithms. In general, all HPO algorithms considered here need to implement two decision making primitives, *searching* and *scheduling*. First, they need to sample new hyperparameter configurations, which often involves some kind of search over the configuration space. Second, for each configuration, an HPO algorithm needs to schedule its evaluation and decide how many resources to allocate for it. Once we start to evaluate a configuration, we will refer to it as a *trial*. We map these decisions to two classes, `HPOSearcher` and `HPOScheduler`. On top of that, we also provide a `HPOTuner` class that executes the optimization process.

This concept of scheduler and searcher is also implemented in popular HPO libraries, such as Syne Tune (Salinas *et al.*, 2022), Ray Tune (Liaw *et al.*, 2018) or Optuna (Akiba *et al.*, 2019).

```
import time
from scipy import stats
from d2l import torch as d2l
```

### 19.2.1 Searcher

Below we define a base class for searchers, which provides a new candidate configuration through the `sample_configuration` function. A simple way to implement this function would be to sample configurations uniformly at random, as we did for random search in Section 19.1. More sophisticated algorithms, such as Bayesian optimization, will make these decisions based on the performance of previous trials. As a result, these algorithms are able to sample more promising candidates over time. We add the `update` function in order to update the history of previous trials, which can then be exploited to improve our sampling distribution.

```
class HPOSearcher(d2l.HyperParameters):  #@save
    def sample_configuration() -> dict:
        raise NotImplementedError
```

```
    def update(self, config: dict, error: float, additional_info=None):
        pass
```

The following code shows how to implement our random search optimizer from the previous section in this API. As a slight extension, we allow the user to prescribe the first configuration to be evaluated via `initial_config`, while subsequent ones are drawn at random.

```
class RandomSearcher(HPOSearcher):  #@save
    def __init__(self, config_space: dict, initial_config=None):
        self.save_hyperparameters()

    def sample_configuration(self) -> dict:
        if self.initial_config is not None:
            result = self.initial_config
            self.initial_config = None
        else:
            result = {
                name: domain.rvs()
                for name, domain in self.config_space.items()
            }
        return result
```

## 19.2.2 Scheduler

Beyond sampling configurations for new trials, we also need to decide when and for how long to run a trial. In practice, all these decisions are done by the `HPOScheduler`, which delegates the choice of new configurations to a `HPOSearcher`. The `suggest` method is called whenever some resource for training becomes available. Apart from invoking `sample_configuration` of a searcher, it may also decide upon parameters like `max_epochs` (i.e., how long to train the model for). The `update` method is called whenever a trial returns a new observation.

```
class HPOScheduler(d2l.HyperParameters):  #@save
    def suggest(self) -> dict:
        raise NotImplementedError

    def update(self, config: dict, error: float, info=None):
        raise NotImplementedError
```

To implement random search, but also other HPO algorithms, we only need a basic scheduler that schedules a new configuration every time new resources become available.

```
class BasicScheduler(HPOScheduler):  #@save
    def __init__(self, searcher: HPOSearcher):
        self.save_hyperparameters()

    def suggest(self) -> dict:
```

```
        return self.searcher.sample_configuration()

    def update(self, config: dict, error: float, info=None):
        self.searcher.update(config, error, additional_info=info)
```

### 19.2.3 Tuner

Finally, we need a component that runs the scheduler/searcher and does some book-keeping of the results. The following code implements a sequential execution of the HPO trials that evaluates one training job after the next and will serve as a basic example. We will later use *Syne Tune* for more scalable distributed HPO cases.

```
class HPOTuner(d2l.HyperParameters):  #@save
    def __init__(self, scheduler: HPOScheduler, objective: callable):
        self.save_hyperparameters()
        # Bookeeping results for plotting
        self.incumbent = None
        self.incumbent_error = None
        self.incumbent_trajectory = []
        self.cumulative_runtime = []
        self.current_runtime = 0
        self.records = []

    def run(self, number_of_trials):
        for i in range(number_of_trials):
            start_time = time.time()
            config = self.scheduler.suggest()
            print(f"Trial {i}: config = {config}")
            error = self.objective(**config)
            error = float(error.cpu().detach().numpy())
            self.scheduler.update(config, error)
            runtime = time.time() - start_time
            self.bookkeeping(config, error, runtime)
            print(f"    error = {error}, runtime = {runtime}")
```

### 19.2.4 Bookkeeping the Performance of HPO Algorithms

With any HPO algorithm, we are mostly interested in the best performing configuration (called *incumbent*) and its validation error after a given wall-clock time. This is why we track `runtime` per iteration, which includes both the time to run an evaluation (call of `objective`) and the time to make a decision (call of `scheduler.suggest`). In the sequel, we will plot `cumulative_runtime` against `incumbent_trajectory` in order to visualize the *any-time performance* of the HPO algorithm defined in terms of `scheduler` (and `searcher`). This allows us to quantify not only how well the configuration found by an optimizer works, but also how quickly an optimizer is able to find it.

```
@d2l.add_to_class(HPOTuner)  #@save
def bookkeeping(self, config: dict, error: float, runtime: float):
    self.records.append({"config": config, "error": error, "runtime": runtime})
    # Check if the last hyperparameter configuration performs better
    # than the incumbent
    if self.incumbent is None or self.incumbent_error > error:
        self.incumbent = config
        self.incumbent_error = error
    # Add current best observed performance to the optimization trajectory
    self.incumbent_trajectory.append(self.incumbent_error)
    # Update runtime
    self.current_runtime += runtime
    self.cumulative_runtime.append(self.current_runtime)
```

### 19.2.5 Example: Optimizing the Hyperparameters of a Convolutional Neural Network

We now use our new implementation of random search to optimize the *batch size* and *learning rate* of the `LeNet` convolutional neural network from Section 7.6. We being by defining the objective function, which will once more be validation error.

```
def hpo_objective_lenet(learning_rate, batch_size, max_epochs=10):  #@save
    model = d2l.LeNet(lr=learning_rate, num_classes=10)
    trainer = d2l.HPOTrainer(max_epochs=max_epochs, num_gpus=1)
    data = d2l.FashionMNIST(batch_size=batch_size)
    model.apply_init([next(iter(data.get_dataloader(True)))[0]], d2l.init_cnn)
    trainer.fit(model=model, data=data)
    validation_error = trainer.validation_error()
    return validation_error
```

We also need to define the configuration space. Moreover, the first configuration to be evaluated is the default setting used in Section 7.6.

```
config_space = {
    "learning_rate": stats.loguniform(1e-2, 1),
    "batch_size": stats.randint(32, 256),
}
initial_config = {
    "learning_rate": 0.1,
    "batch_size": 128,
}
```

Now we can start our random search:

```
searcher = RandomSearcher(config_space, initial_config=initial_config)
scheduler = BasicScheduler(searcher=searcher)
tuner = HPOTuner(scheduler=scheduler, objective=hpo_objective_lenet)
tuner.run(number_of_trials=5)
```

```
error = 0.47701597213745117, runtime = 90.8468370437622
```

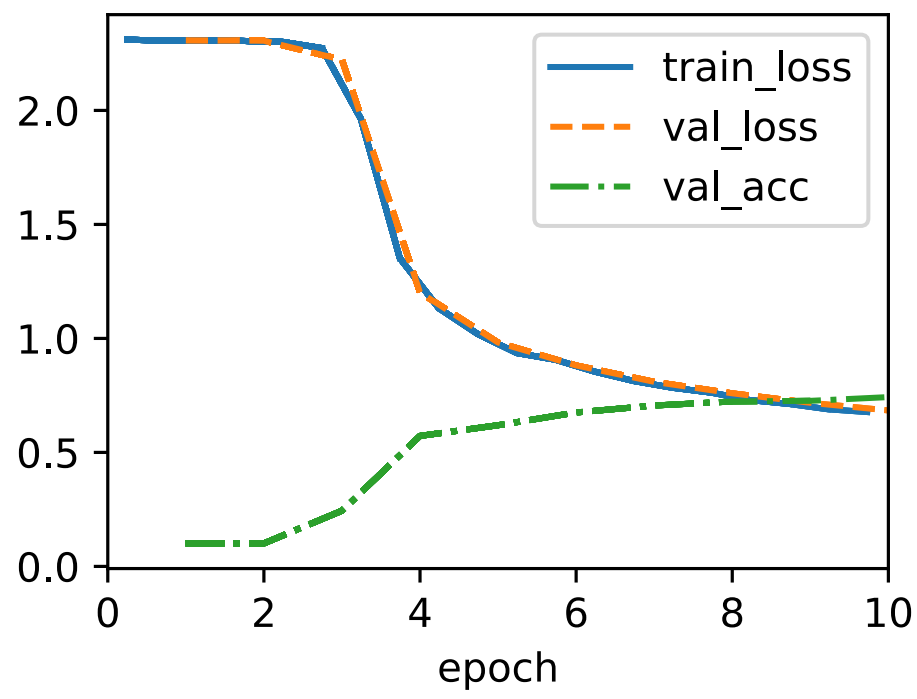

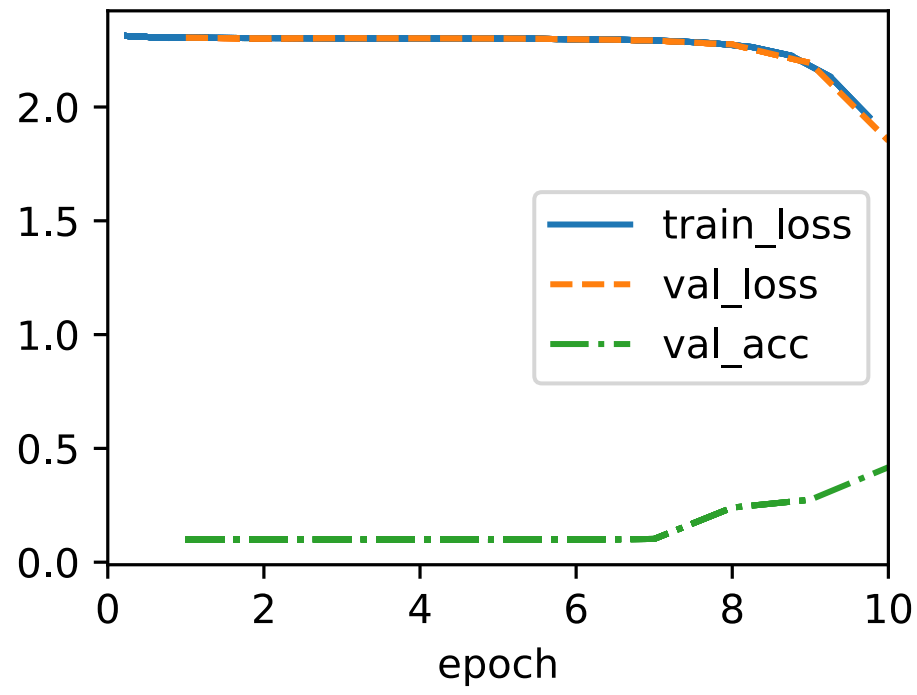

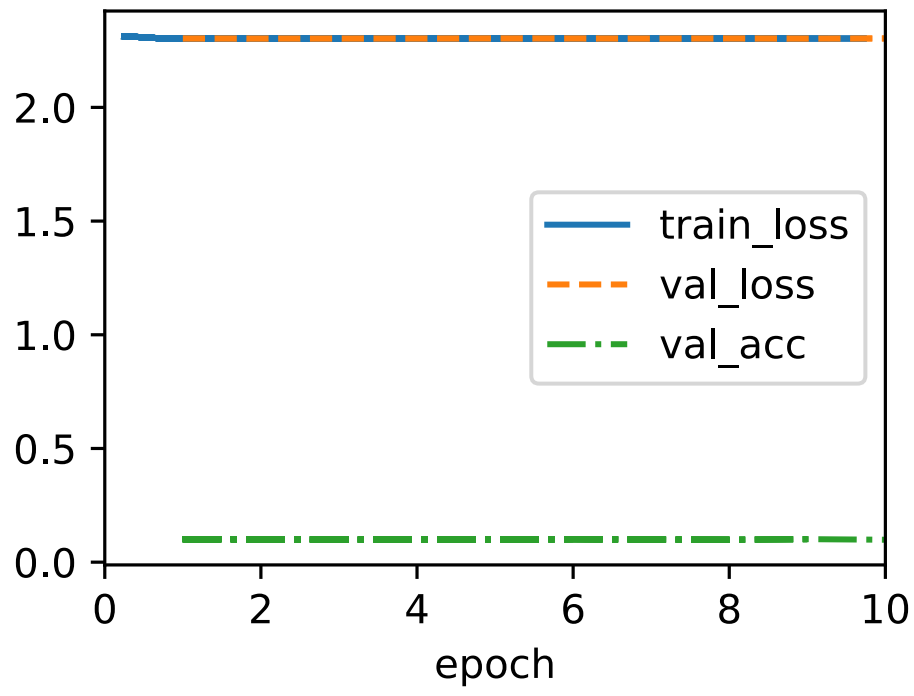

Below we plot the optimization trajectory of the incumbent to get the any-time performance of random search:

```
board = d2l.ProgressBoard(xlabel="time", ylabel="error")
for time_stamp, error in zip(
    tuner.cumulative_runtime, tuner.incumbent_trajectory
```

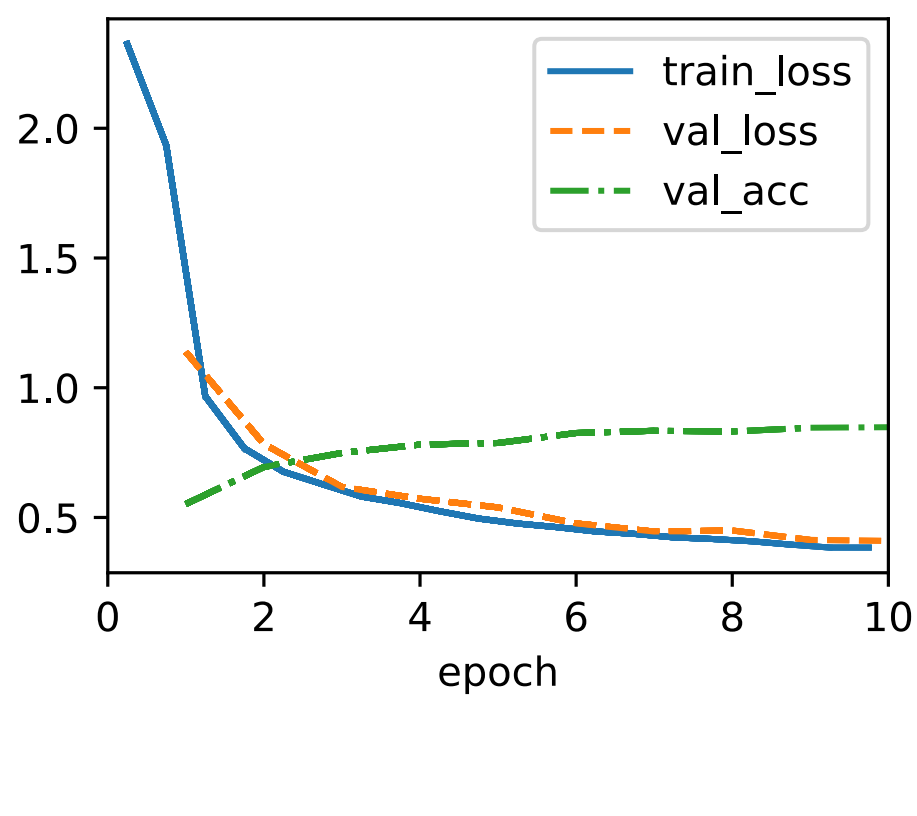

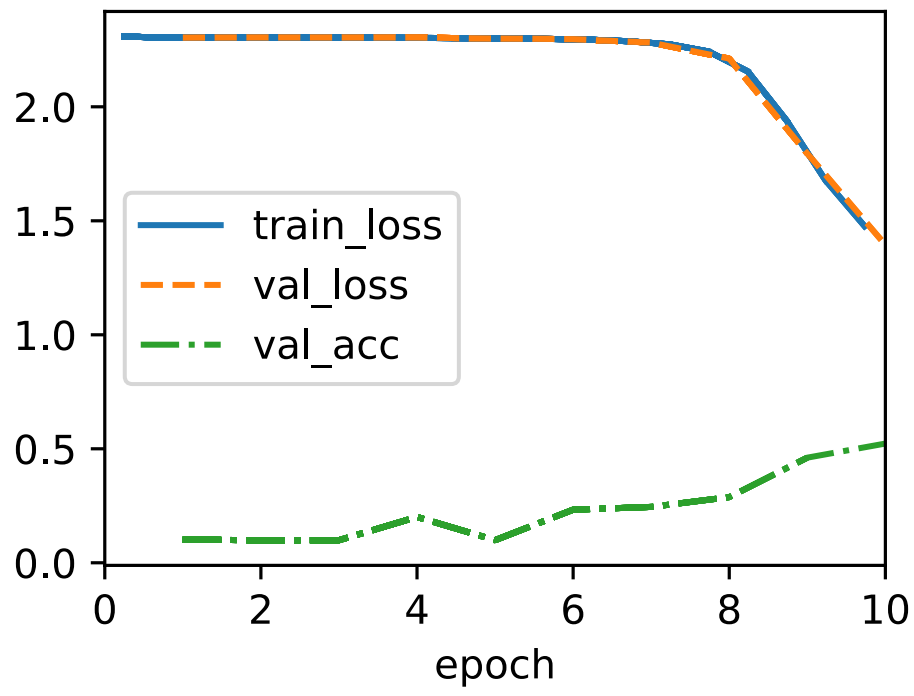

```
):
    board.draw(time_stamp, error, "random search", every_n=1)
```

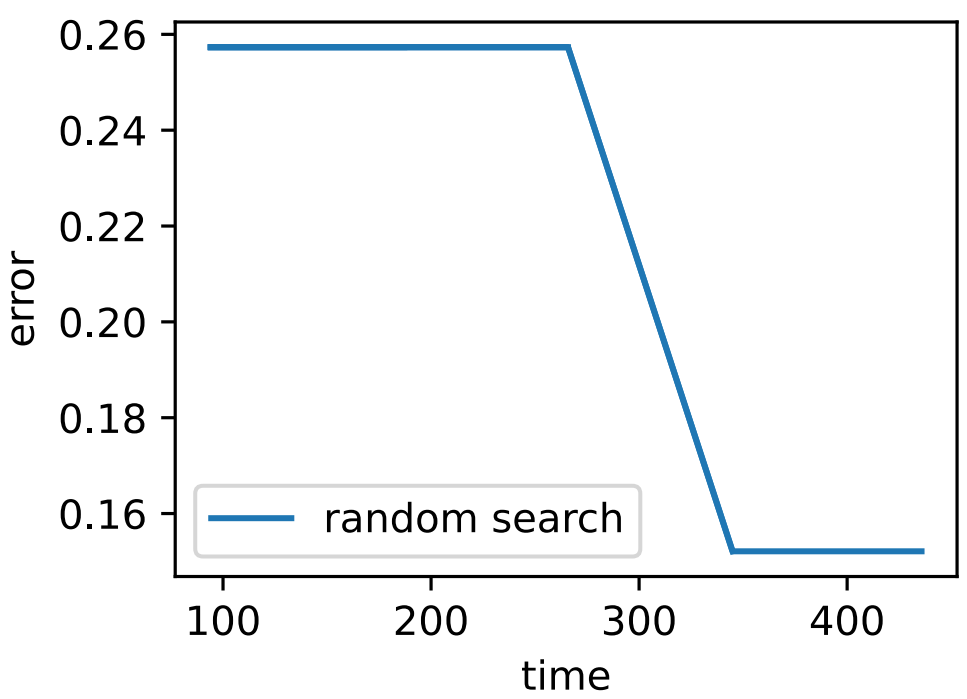

### 19.2.6 Comparing HPO Algorithms

Just as with training algorithms or model architectures, it is important to understand how to best compare different HPO algorithms. Each HPO run depends on two major sources of randomness: the random effects of the training process, such as random weight initialization or mini-batch ordering, and the intrinsic randomness of the HPO algorithm itself, such as the random sampling of random search. Hence, when comparing different algorithms, it is crucial to run each experiment several times and report statistics, such as mean or median, across a population of multiple repetitions of an algorithm based on different seeds of the random number generator.

To illustrate this, we compare random search (see Section 19.1.2) and Bayesian optimization (Snoek *et al.*, 2012) on tuning the hyperparameters of a feed-forward neural network. Each algorithm was evaluated 50 times with a different random seed. The solid line indicates the average performance of the incumbent across these 50 repetitions and the dashed line the standard deviation. We can see that random search and Bayesian optimization perform roughly the same up to ~1000 seconds, but Bayesian optimization can make use of the past observation to identify better configurations and thus quickly outperforms random search afterwards.

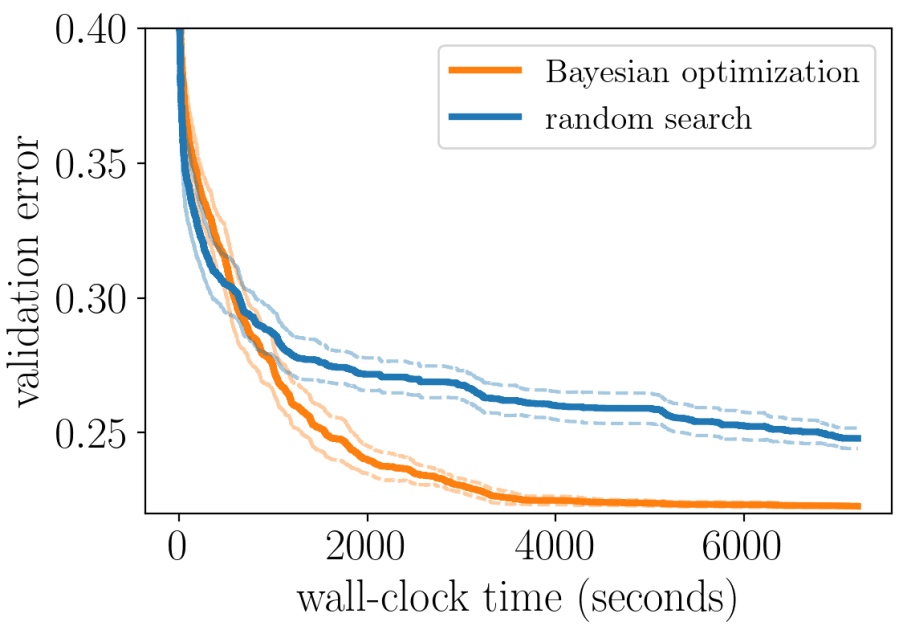

Fig. 19.2.1 Example any-time performance plot to compare two algorithms A and B.

## 19.2.7 Summary

This section laid out a simple, yet flexible interface to implement various HPO algorithms that we will look at in this chapter. Similar interfaces can be found in popular open-source HPO frameworks. We also looked at how we can compare HPO algorithms, and potential pitfall one needs to be aware.

## 19.2.8 Exercises

1. The goal of this exercise is to implement the objective function for a slightly more challenging HPO problem, and to run more realistic experiments. We will use the two hidden layer MLP `DropoutMLP` implemented in Section 5.6.

    1. Code up the objective function, which should depend on all hyperparameters of the model and `batch_size`. Use `max_epochs=50`. GPUs do not help here, so `num_gpus=0`. Hint: Modify `hpo_objective_lenet`.

2. Choose a sensible search space, where `num_hiddens_1`, `num_hiddens_2` are integers in $[8, 1024]$, and dropout values lie in $[0, 0.95]$, while `batch_size` lies in $[16, 384]$. Provide code for `config_space`, using sensible distributions from `scipy.stats`.

3. Run random search on this example with `number_of_trials=20` and plot the results. Make sure to first evaluate the default configuration of Section 5.6, which is `initial_config = {'num_hiddens_1': 256, 'num_hiddens_2': 256, 'dropout_1': 0.5, 'dropout_2': 0.5, 'lr': 0.1, 'batch_size': 256}`.

2. In this exercise, you will implement a new searcher (subclass of `HPOSearcher`) which makes decisions based on past data. It depends on parameters `probab_local`, `num_init_random`. Its `sample_configuration` method works as follows. For the first `num_init_random` calls, do the same as `RandomSearcher.sample_configuration`. Otherwise, with probability `1 - probab_local`, do the same as `RandomSearcher.sample_configuration`. Otherwise, pick the configuration which attained the smallest validation error so far, select one of its hyperparameters at random, and sample its value randomly like in `RandomSearcher.sample_configuration`, but leave all other values the same. Return this configuration, which is identical to the best configuration so far, except in this one hyperparameter.

1. Code up this new `LocalSearcher`. Hint: Your searcher requires `config_space` as argument at construction. Feel free to use a member of type `RandomSearcher`. You will also have to implement the `update` method.

2. Re-run the experiment from the previous exercise, but using your new searcher instead of `RandomSearcher`. Experiment with different values for `probab_local`, `num_init_random`. However, note that a proper comparison between different HPO methods requires repeating experiments several times, and ideally considering a number of benchmark tasks.

Discussions[266].

266

## 19.3 Asynchronous Random Search

As we have seen in the previous Section 19.2, we might have to wait hours or even days before random search returns a good hyperparameter configuration, because of the expensive evaluation of hyperparameter configurations. In practice, we have often access to a pool of resources such as multiple GPUs on the same machine or multiple machines with a single GPU. This begs the question: *How do we efficiently distribute random search?*

In general, we distinguish between synchronous and asynchronous parallel hyperparameter optimization (see Fig. 19.3.1). In the synchronous setting, we wait for all concurrently running trials to finish, before we start the next batch. Consider configuration spaces that contain hyperparameters such as the number of filters or number of layers of a deep neural network. Hyperparameter configurations that contain a larger number of layers of filters will naturally take more time to finish, and all other trials in the same batch will have to wait at synchronisation points (grey area in Fig. 19.3.1) before we can continue the optimization process.

In the asynchronous setting we immediately schedule a new trial as soon as resources become available. This will optimally exploit our resources, since we can avoid any synchronisation overhead. For random search, each new hyperparameter configuration is chosen independently of all others, and in particular without exploiting observations from any prior evaluation. This means we can trivially parallelize random search asynchronously. This is not straight-forward with more sophisticated methods that make decision based on previous observations (see Section 19.5). While we need access to more resources than in the sequential setting, asynchronous random search exhibits a linear speed-up, in that a certain performance is reached $K$ times faster if $K$ trials can be run in parallel.

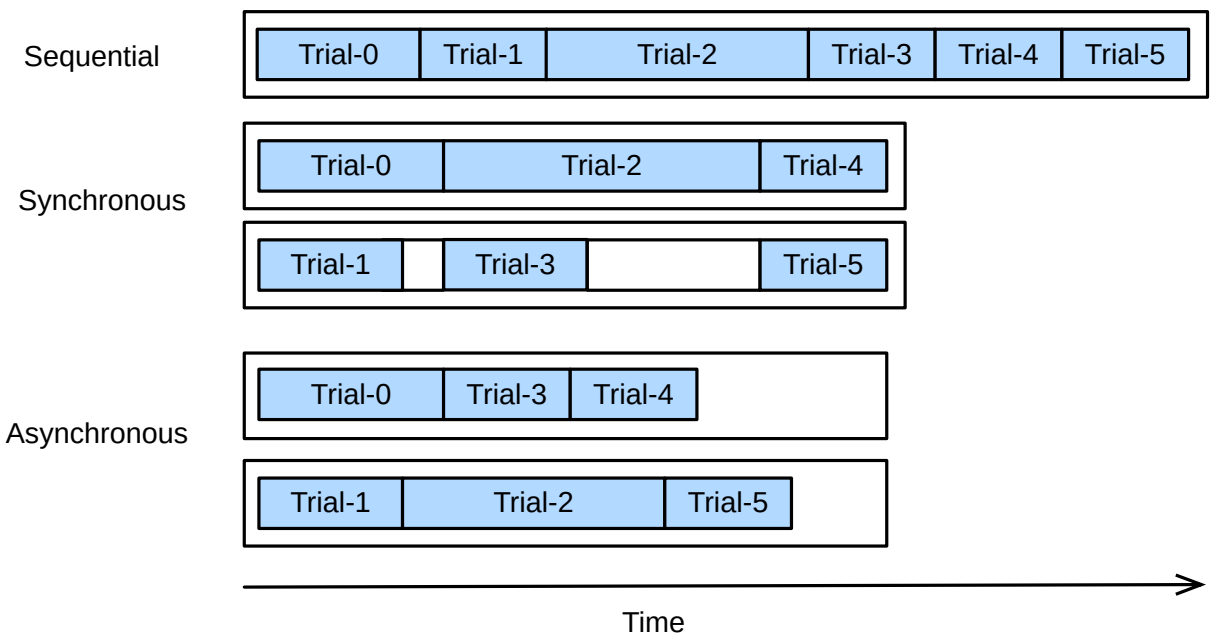

Fig. 19.3.1 Distributing the hyperparameter optimization process either synchronously or asynchronously. Compared to the sequential setting, we can reduce the overall wall-clock time while keep the total compute constant. Synchronous scheduling might lead to idling workers in the case of stragglers.

In this notebook, we will look at asynchronous random search that, where trials are executed in multiple python processes on the same machine. Distributed job scheduling and execution is difficult to implement from scratch. We will use *Syne Tune* (Salinas *et al.*, 2022), which provides us with a simple interface for asynchronous HPO. Syne Tune is designed to be run with different execution back-ends, and the interested reader is invited to study its simple APIs in order to learn more about distributed HPO.

```
import logging
from d2l import torch as d2l

logging.basicConfig(level=logging.INFO)
from syne_tune import StoppingCriterion, Tuner
from syne_tune.backend.python_backend import PythonBackend
from syne_tune.config_space import loguniform, randint
from syne_tune.experiments import load_experiment
from syne_tune.optimizer.baselines import RandomSearch
```

```
INFO:root:SageMakerBackend is not imported since dependencies are missing. You␣
```

```
↪can install them with
   pip install 'syne-tune[extra]'
AWS dependencies are not imported since dependencies are missing. You can␣
↪install them with
   pip install 'syne-tune[aws]'
or (for everything)
   pip install 'syne-tune[extra]'
AWS dependencies are not imported since dependencies are missing. You can␣
↪install them with
   pip install 'syne-tune[aws]'
or (for everything)
   pip install 'syne-tune[extra]'
INFO:root:Ray Tune schedulers and searchers are not imported since␣
↪dependencies are missing. You can install them with
   pip install 'syne-tune[raytune]'
or (for everything)
   pip install 'syne-tune[extra]'
```

## 19.3.1 Objective Function

First, we have to define a new objective function such that it now returns the performance back to Syne Tune via the `report` callback.

```python
def hpo_objective_lenet_synetune(learning_rate, batch_size, max_epochs):
    from syne_tune import Reporter
    from d2l import torch as d2l

    model = d2l.LeNet(lr=learning_rate, num_classes=10)
    trainer = d2l.HPOTrainer(max_epochs=1, num_gpus=1)
    data = d2l.FashionMNIST(batch_size=batch_size)
    model.apply_init([next(iter(data.get_dataloader(True)))[0]], d2l.init_cnn)
    report = Reporter()
    for epoch in range(1, max_epochs + 1):
        if epoch == 1:
            # Initialize the state of Trainer
            trainer.fit(model=model, data=data)
        else:
            trainer.fit_epoch()
        validation_error = trainer.validation_error().cpu().detach().numpy()
        report(epoch=epoch, validation_error=float(validation_error))
```

Note that the `PythonBackend` of Syne Tune requires dependencies to be imported inside the function definition.

## 19.3.2 Asynchronous Scheduler

First, we define the number of workers that evaluate trials concurrently. We also need to specify how long we want to run random search, by defining an upper limit on the total wall-clock time.

```
n_workers = 2  # Needs to be <= the number of available GPUs

max_wallclock_time = 12 * 60  # 12 minutes
```

Next, we state which metric we want to optimize and whether we want to minimize or maximize this metric. Namely, `metric` needs to correspond to the argument name passed to the `report` callback.

```
mode = "min"
metric = "validation_error"
```

We use the configuration space from our previous example. In Syne Tune, this dictionary can also be used to pass constant attributes to the training script. We make use of this feature in order to pass `max_epochs`. Moreover, we specify the first configuration to be evaluated in `initial_config`.

```
config_space = {
    "learning_rate": loguniform(1e-2, 1),
    "batch_size": randint(32, 256),
    "max_epochs": 10,
}
initial_config = {
    "learning_rate": 0.1,
    "batch_size": 128,
}
```

Next, we need to specify the back-end for job executions. Here we just consider the distribution on a local machine where parallel jobs are executed as sub-processes. However, for large scale HPO, we could run this also on a cluster or cloud environment, where each trial consumes a full instance.

```
trial_backend = PythonBackend(
    tune_function=hpo_objective_lenet_synetune,
    config_space=config_space,
)
```

We can now create the scheduler for asynchronous random search, which is similar in behaviour to our `BasicScheduler` from Section 19.2.

```
scheduler = RandomSearch(
    config_space,
    metric=metric,
    mode=mode,
    points_to_evaluate=[initial_config],
)
```

```
INFO:syne_tune.optimizer.schedulers.fifo:max_resource_level = 10, as inferred␣
```

```
↪from config_space
INFO:syne_tune.optimizer.schedulers.fifo:Master random_seed = 218748764
```

Syne Tune also features a Tuner, where the main experiment loop and bookkeeping is centralized, and interactions between scheduler and back-end are mediated.

```
stop_criterion = StoppingCriterion(max_wallclock_time=max_wallclock_time)

tuner = Tuner(
    trial_backend=trial_backend,
    scheduler=scheduler,
    stop_criterion=stop_criterion,
    n_workers=n_workers,
    print_update_interval=int(max_wallclock_time * 0.6),
)
```

Let us run our distributed HPO experiment. According to our stopping criterion, it will run for about 12 minutes.

```
tuner.run()
```

```
INFO:syne_tune.tuner:results of trials will be saved on /home/ubuntu/syne-tune/
↪python-entrypoint-2023-08-18-23-04-42-649
INFO:root:Detected 8 GPUs
INFO:root:running subprocess with command: /home/ubuntu/miniconda3/envs/np3/
↪bin/python /home/ubuntu/miniconda3/envs/np3/lib/python3.9/site-packages/syne_
↪tune/backend/python_backend/python_entrypoint.py --learning_rate 0.1 --batch_
↪size 128 --max_epochs 10 --tune_function_root /home/ubuntu/syne-tune/python-
↪entrypoint-2023-08-18-23-04-42-649/tune_function --tune_function_hash␣
↪c3e045478958babaa9af3aa66092328d --st_checkpoint_dir /home/ubuntu/syne-tune/
↪python-entrypoint-2023-08-18-23-04-42-649/0/checkpoints
INFO:syne_tune.tuner:(trial 0) - scheduled config {'learning_rate': 0.1,
↪'batch_size': 128, 'max_epochs': 10}
INFO:root:running subprocess with command: /home/ubuntu/miniconda3/envs/np3/
↪bin/python /home/ubuntu/miniconda3/envs/np3/lib/python3.9/site-packages/syne_
↪tune/backend/python_backend/python_entrypoint.py --learning_rate 0.
↪11573821391982765 --batch_size 249 --max_epochs 10 --tune_function_root /
↪home/ubuntu/syne-tune/python-entrypoint-2023-08-18-23-04-42-649/tune_
↪function --tune_function_hash c3e045478958babaa9af3aa66092328d --st_
↪checkpoint_dir /home/ubuntu/syne-tune/python-entrypoint-2023-08-18-23-04-42-
↪649/1/checkpoints
INFO:syne_tune.tuner:(trial 1) - scheduled config {'learning_rate': 0.
↪11573821391982765, 'batch_size': 249, 'max_epochs': 10}
INFO:syne_tune.tuner:Trial trial_id 0 completed.
INFO:syne_tune.tuner:Trial trial_id 1 completed.
INFO:root:running subprocess with command: /home/ubuntu/miniconda3/envs/np3/
↪bin/python /home/ubuntu/miniconda3/envs/np3/lib/python3.9/site-packages/syne_
↪tune/backend/python_backend/python_entrypoint.py --learning_rate 0.
↪1270190537228491 --batch_size 180 --max_epochs 10 --tune_function_root /home/
↪ubuntu/syne-tune/python-entrypoint-2023-08-18-23-04-42-649/tune_function --
↪tune_function_hash c3e045478958babaa9af3aa66092328d --st_checkpoint_dir /
```

```
↪home/ubuntu/syne-tune/python-entrypoint-2023-08-18-23-04-42-649/2/checkpoints
INFO:syne_tune.tuner:(trial 2) - scheduled config {'learning_rate': 0.
↪1270190537228491, 'batch_size': 180, 'max_epochs': 10}
INFO:root:running subprocess with command: /home/ubuntu/miniconda3/envs/np3/
↪bin/python /home/ubuntu/miniconda3/envs/np3/lib/python3.9/site-packages/syne_
↪tune/backend/python_backend/python_entrypoint.py --learning_rate 0.
↪013539985569129874 --batch_size 44 --max_epochs 10 --tune_function_root /
↪home/ubuntu/syne-tune/python-entrypoint-2023-08-18-23-04-42-649/tune_
↪function --tune_function_hash c3e045478958babaa9af3aa66092328d --st_
↪checkpoint_dir /home/ubuntu/syne-tune/python-entrypoint-2023-08-18-23-04-42-
↪649/3/checkpoints
INFO:syne_tune.tuner:(trial 3) - scheduled config {'learning_rate': 0.
↪013539985569129874, 'batch_size': 44, 'max_epochs': 10}
INFO:syne_tune.tuner:Trial trial_id 2 completed.
INFO:root:running subprocess with command: /home/ubuntu/miniconda3/envs/np3/
↪bin/python /home/ubuntu/miniconda3/envs/np3/lib/python3.9/site-packages/syne_
↪tune/backend/python_backend/python_entrypoint.py --learning_rate 0.
↪5623908565974269 --batch_size 117 --max_epochs 10 --tune_function_root /home/
↪ubuntu/syne-tune/python-entrypoint-2023-08-18-23-04-42-649/tune_function --
↪tune_function_hash c3e045478958babaa9af3aa66092328d --st_checkpoint_dir /
↪home/ubuntu/syne-tune/python-entrypoint-2023-08-18-23-04-42-649/4/checkpoints
INFO:syne_tune.tuner:(trial 4) - scheduled config {'learning_rate': 0.
↪5623908565974269, 'batch_size': 117, 'max_epochs': 10}
INFO:syne_tune.tuner:Trial trial_id 3 completed.
INFO:root:running subprocess with command: /home/ubuntu/miniconda3/envs/np3/
↪bin/python /home/ubuntu/miniconda3/envs/np3/lib/python3.9/site-packages/syne_
↪tune/backend/python_backend/python_entrypoint.py --learning_rate 0.
↪4237481357205294 --batch_size 217 --max_epochs 10 --tune_function_root /home/
↪ubuntu/syne-tune/python-entrypoint-2023-08-18-23-04-42-649/tune_function --
↪tune_function_hash c3e045478958babaa9af3aa66092328d --st_checkpoint_dir /
↪home/ubuntu/syne-tune/python-entrypoint-2023-08-18-23-04-42-649/5/checkpoints
INFO:syne_tune.tuner:(trial 5) - scheduled config {'learning_rate': 0.
↪4237481357205294, 'batch_size': 217, 'max_epochs': 10}
INFO:syne_tune.tuner:Trial trial_id 4 completed.
INFO:root:running subprocess with command: /home/ubuntu/miniconda3/envs/np3/
↪bin/python /home/ubuntu/miniconda3/envs/np3/lib/python3.9/site-packages/syne_
↪tune/backend/python_backend/python_entrypoint.py --learning_rate 0.
↪14543999948294017 --batch_size 100 --max_epochs 10 --tune_function_root /
↪home/ubuntu/syne-tune/python-entrypoint-2023-08-18-23-04-42-649/tune_
↪function --tune_function_hash c3e045478958babaa9af3aa66092328d --st_
↪checkpoint_dir /home/ubuntu/syne-tune/python-entrypoint-2023-08-18-23-04-42-
↪649/6/checkpoints
INFO:syne_tune.tuner:(trial 6) - scheduled config {'learning_rate': 0.
↪14543999948294017, 'batch_size': 100, 'max_epochs': 10}
INFO:syne_tune.tuner:Trial trial_id 5 completed.
INFO:root:running subprocess with command: /home/ubuntu/miniconda3/envs/np3/
↪bin/python /home/ubuntu/miniconda3/envs/np3/lib/python3.9/site-packages/syne_
↪tune/backend/python_backend/python_entrypoint.py --learning_rate 0.
↪013237796900392465 --batch_size 184 --max_epochs 10 --tune_function_root /
↪home/ubuntu/syne-tune/python-entrypoint-2023-08-18-23-04-42-649/tune_
↪function --tune_function_hash c3e045478958babaa9af3aa66092328d --st_
↪checkpoint_dir /home/ubuntu/syne-tune/python-entrypoint-2023-08-18-23-04-42-
↪649/7/checkpoints
INFO:syne_tune.tuner:(trial 7) - scheduled config {'learning_rate': 0.
↪013237796900392465, 'batch_size': 184, 'max_epochs': 10}
```

```
INFO:syne_tune.tuner:Trial trial_id 6 completed.
INFO:root:running subprocess with command: /home/ubuntu/miniconda3/envs/np3/
↪bin/python /home/ubuntu/miniconda3/envs/np3/lib/python3.9/site-packages/syne_
↪tune/backend/python_backend/python_entrypoint.py --learning_rate 0.
↪019419074864676922 --batch_size 147 --max_epochs 10 --tune_function_root /
↪home/ubuntu/syne-tune/python-entrypoint-2023-08-18-23-04-42-649/tune_
↪function --tune_function_hash c3e045478958babaa9af3aa66092328d --st_
↪checkpoint_dir /home/ubuntu/syne-tune/python-entrypoint-2023-08-18-23-04-42-
↪649/8/checkpoints
INFO:syne_tune.tuner:(trial 8) - scheduled config {'learning_rate': 0.
↪019419074864676922, 'batch_size': 147, 'max_epochs': 10}
INFO:syne_tune.tuner:tuning status (last metric is reported)
 trial_id     status  iter  learning_rate  batch_size  max_epochs  epoch ␣
↪validation_error  worker-time
        0  Completed    10       0.100000         128          10   10.0      ␣
↪   0.268987    98.142261
        1  Completed    10       0.115738         249          10   10.0      ␣
↪   0.380865    94.529288
        2  Completed    10       0.127019         180          10   10.0      ␣
↪   0.275060    92.739022
        3  Completed    10       0.013540          44          10   10.0      ␣
↪   0.421119   128.757522
        4  Completed    10       0.562391         117          10   10.0      ␣
↪   0.155812    97.749136
        5  Completed    10       0.423748         217          10   10.0      ␣
↪   0.234740    95.035502
        6  Completed    10       0.145440         100          10   10.0      ␣
↪   0.227300    97.266276
        7 InProgress     9       0.013238         184          10    9.0      ␣
↪   0.900074    79.913983
        8 InProgress     0       0.019419         147          10      -      ␣
↪          -            -
2 trials running, 7 finished (7 until the end), 436.49s wallclock-time

INFO:syne_tune.tuner:Trial trial_id 7 completed.
INFO:root:running subprocess with command: /home/ubuntu/miniconda3/envs/np3/
↪bin/python /home/ubuntu/miniconda3/envs/np3/lib/python3.9/site-packages/syne_
↪tune/backend/python_backend/python_entrypoint.py --learning_rate 0.
↪10014301952595842 --batch_size 101 --max_epochs 10 --tune_function_root /
↪home/ubuntu/syne-tune/python-entrypoint-2023-08-18-23-04-42-649/tune_
↪function --tune_function_hash c3e045478958babaa9af3aa66092328d --st_
↪checkpoint_dir /home/ubuntu/syne-tune/python-entrypoint-2023-08-18-23-04-42-
↪649/9/checkpoints
INFO:syne_tune.tuner:(trial 9) - scheduled config {'learning_rate': 0.
↪10014301952595842, 'batch_size': 101, 'max_epochs': 10}
INFO:syne_tune.tuner:Trial trial_id 8 completed.
INFO:root:running subprocess with command: /home/ubuntu/miniconda3/envs/np3/
↪bin/python /home/ubuntu/miniconda3/envs/np3/lib/python3.9/site-packages/syne_
↪tune/backend/python_backend/python_entrypoint.py --learning_rate 0.
↪011852303154416318 --batch_size 191 --max_epochs 10 --tune_function_root /
↪home/ubuntu/syne-tune/python-entrypoint-2023-08-18-23-04-42-649/tune_
↪function --tune_function_hash c3e045478958babaa9af3aa66092328d --st_
↪checkpoint_dir /home/ubuntu/syne-tune/python-entrypoint-2023-08-18-23-04-42-
↪649/10/checkpoints
INFO:syne_tune.tuner:(trial 10) - scheduled config {'learning_rate': 0.
```

```
↪011852303154416318, 'batch_size': 191, 'max_epochs': 10}
INFO:syne_tune.tuner:Trial trial_id 9 completed.
INFO:root:running subprocess with command: /home/ubuntu/miniconda3/envs/np3/
↪bin/python /home/ubuntu/miniconda3/envs/np3/lib/python3.9/site-packages/syne_
↪tune/backend/python_backend/python_entrypoint.py --learning_rate 0.
↪04352474626996822 --batch_size 45 --max_epochs 10 --tune_function_root /home/
↪ubuntu/syne-tune/python-entrypoint-2023-08-18-23-04-42-649/tune_function --
↪tune_function_hash c3e045478958babaa9af3aa66092328d --st_checkpoint_dir /
↪home/ubuntu/syne-tune/python-entrypoint-2023-08-18-23-04-42-649/11/
↪checkpoints
INFO:syne_tune.tuner:(trial 11) - scheduled config {'learning_rate': 0.
↪04352474626996822, 'batch_size': 45, 'max_epochs': 10}
INFO:syne_tune.tuner:Trial trial_id 10 completed.
INFO:root:running subprocess with command: /home/ubuntu/miniconda3/envs/np3/
↪bin/python /home/ubuntu/miniconda3/envs/np3/lib/python3.9/site-packages/syne_
↪tune/backend/python_backend/python_entrypoint.py --learning_rate 0.
↪6523407101123465 --batch_size 240 --max_epochs 10 --tune_function_root /home/
↪ubuntu/syne-tune/python-entrypoint-2023-08-18-23-04-42-649/tune_function --
↪tune_function_hash c3e045478958babaa9af3aa66092328d --st_checkpoint_dir /
↪home/ubuntu/syne-tune/python-entrypoint-2023-08-18-23-04-42-649/12/
↪checkpoints
INFO:syne_tune.tuner:(trial 12) - scheduled config {'learning_rate': 0.
↪6523407101123465, 'batch_size': 240, 'max_epochs': 10}
INFO:syne_tune.tuner:Trial trial_id 12 completed.
INFO:root:running subprocess with command: /home/ubuntu/miniconda3/envs/np3/
↪bin/python /home/ubuntu/miniconda3/envs/np3/lib/python3.9/site-packages/syne_
↪tune/backend/python_backend/python_entrypoint.py --learning_rate 0.
↪0431019557345446 --batch_size 73 --max_epochs 10 --tune_function_root /home/
↪ubuntu/syne-tune/python-entrypoint-2023-08-18-23-04-42-649/tune_function --
↪tune_function_hash c3e045478958babaa9af3aa66092328d --st_checkpoint_dir /
↪home/ubuntu/syne-tune/python-entrypoint-2023-08-18-23-04-42-649/13/
↪checkpoints
INFO:syne_tune.tuner:(trial 13) - scheduled config {'learning_rate': 0.
↪0431019557345446, 'batch_size': 73, 'max_epochs': 10}
INFO:syne_tune.stopping_criterion:reaching max wallclock time (720), stopping␣
↪there.
INFO:syne_tune.tuner:Stopping trials that may still be running.
INFO:syne_tune.tuner:Tuning finished, results of trials can be found on /home/
↪ubuntu/syne-tune/python-entrypoint-2023-08-18-23-04-42-649
--------------------
Resource summary (last result is reported):
 trial_id     status  iter  learning_rate  batch_size  max_epochs  epoch ␣
↪validation_error  worker-time
        0  Completed    10       0.100000         128          10   10.0      ␣
↪   0.268987     98.142261
        1  Completed    10       0.115738         249          10   10.0      ␣
↪   0.380865     94.529288
        2  Completed    10       0.127019         180          10   10.0      ␣
↪   0.275060     92.739022
        3  Completed    10       0.013540          44          10   10.0      ␣
↪   0.421119    128.757522
        4  Completed    10       0.562391         117          10   10.0      ␣
↪   0.155812     97.749136
        5  Completed    10       0.423748         217          10   10.0      ␣
↪   0.234740     95.035502
```

```
     6  Completed    10       0.145440        100          10   10.0       ␣
↪  0.227300    97.266276
     7  Completed    10       0.013238        184          10   10.0       ␣
↪  0.900086    89.168077
     8  Completed    10       0.019419        147          10   10.0       ␣
↪  0.901410    96.518268
     9  Completed    10       0.100143        101          10   10.0       ␣
↪  0.280792    96.687997
    10  Completed    10       0.011852        191          10   10.0       ␣
↪  0.900143    90.441659
    11 InProgress     9       0.043525         45          10    9.0       ␣
↪  0.264026   144.953945
    12  Completed    10       0.652341        240          10   10.0       ␣
↪  0.185565    86.874975
    13 InProgress     0       0.043102         73          10      -       ␣
↪         -            -
2 trials running, 12 finished (12 until the end), 722.77s wallclock-time

validation_error: best 0.15581202507019043 for trial-id 4
--------------------
```

The logs of all evaluated hyperparameter configurations are stored for further analysis. At any time during the tuning job, we can easily get the results obtained so far and plot the incumbent trajectory.

```
d2l.set_figsize()
tuning_experiment = load_experiment(tuner.name)
tuning_experiment.plot()
```

```
WARNING:matplotlib.legend:No artists with labels found to put in legend.  Note␣
↪that artists whose label start with an underscore are ignored when legend()␣
↪is called with no argument.
```

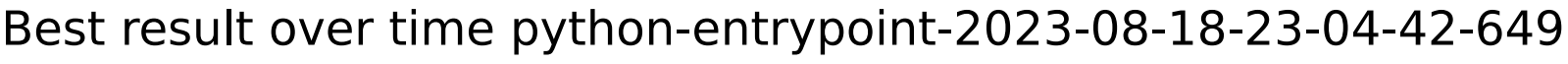

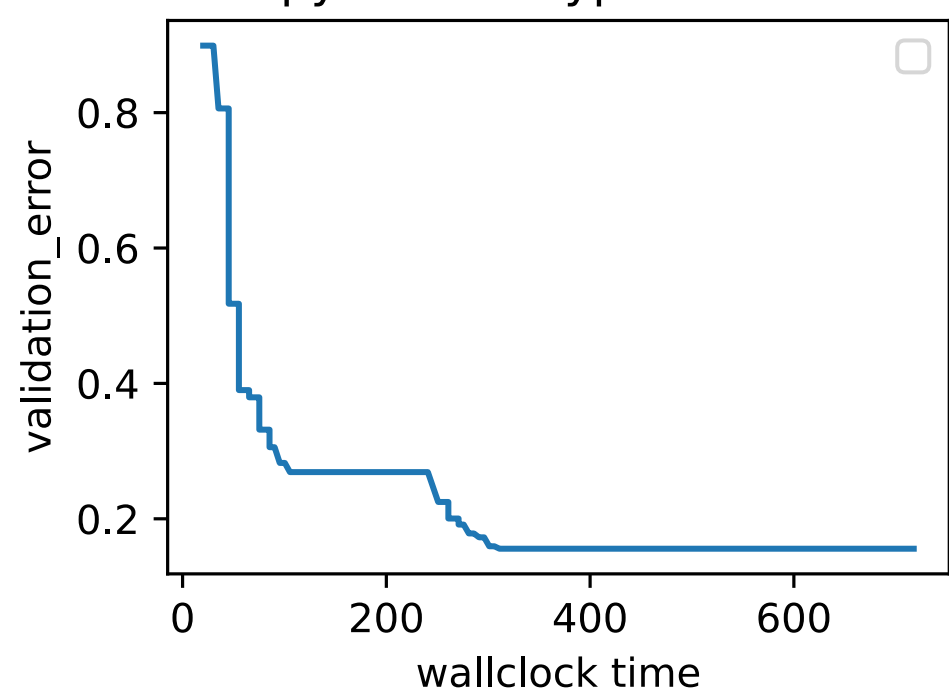

## 19.3.3 Visualize the Asynchronous Optimization Process

Below we visualize how the learning curves of every trial (each color in the plot represents a trial) evolve during the asynchronous optimization process. At any point in time, there are as many trials running concurrently as we have workers. Once a trial finishes, we immediately start the next trial, without waiting for the other trials to finish. Idle time of workers is reduced to a minimum with asynchronous scheduling.

```
d2l.set_figsize([6, 2.5])
results = tuning_experiment.results

for trial_id in results.trial_id.unique():
    df = results[results["trial_id"] == trial_id]
    d2l.plt.plot(
        df["st_tuner_time"],
        df["validation_error"],
        marker="o"
    )

d2l.plt.xlabel("wall-clock time")
d2l.plt.ylabel("objective function")
```

```
Text(0, 0.5, 'objective function')
```

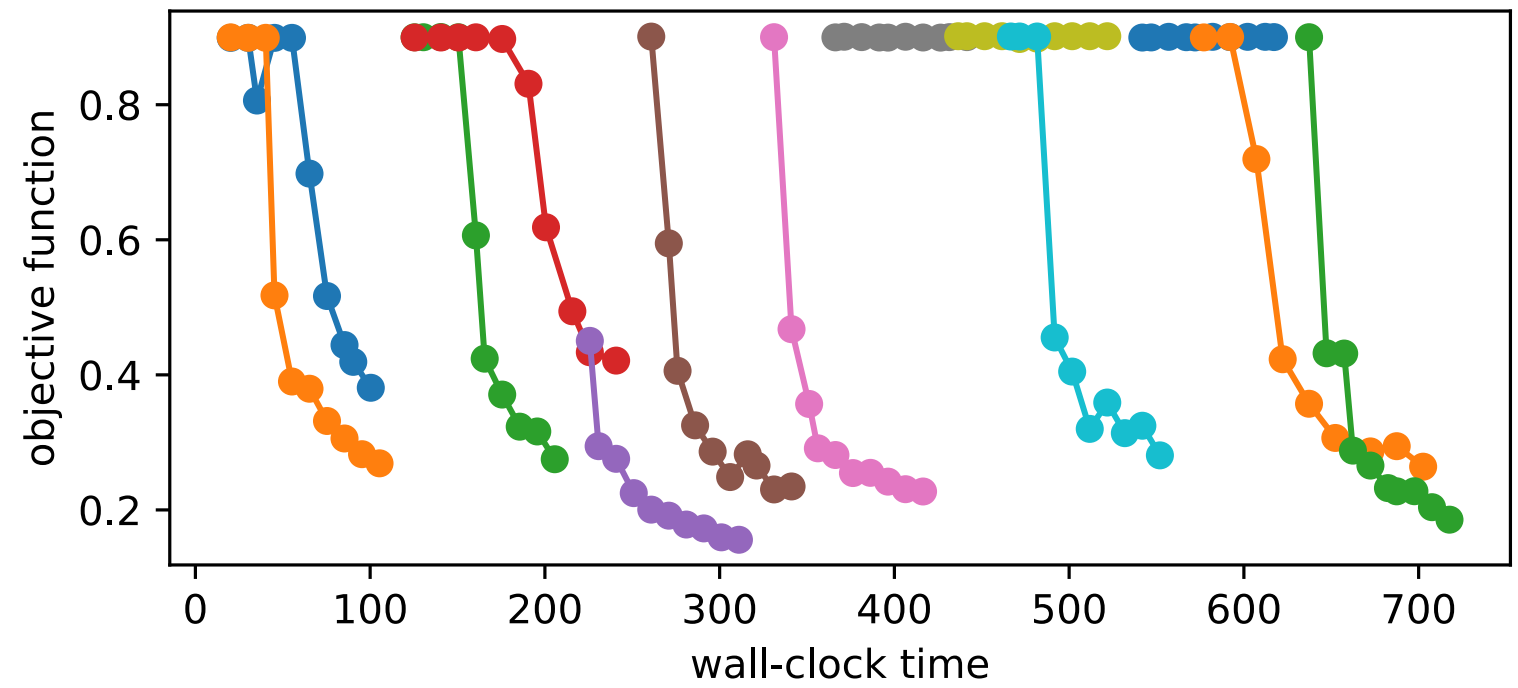

## 19.3.4 Summary

We can reduce the waiting time for random search substantially by distribution trials across parallel resources. In general, we distinguish between synchronous scheduling and asynchronous scheduling. Synchronous scheduling means that we sample a new batch of hyperparameter configurations once the previous batch finished. If we have a stragglers - trials that takes more time to finish than other trials - our workers need to wait at synchronization points. Asynchronous scheduling evaluates a new hyperparameter configurations as soon as resources become available, and, hence, ensures that all workers are busy at any point in time. While random search is easy to distribute asynchronously and does not require any change of the actual algorithm, other methods require some additional modifications.

### 19.3.5 Exercises

1. Consider the `DropoutMLP` model implemented in Section 5.6, and used in Exercise 1 of Section 19.2.

    1. Implement an objective function `hpo_objective_dropoutmlp_synetune` to be used with Syne Tune. Make sure that your function reports the validation error after every epoch.

    2. Using the setup of Exercise 1 in Section 19.2, compare random search to Bayesian optimization. If you use SageMaker, feel free to use Syne Tune's benchmarking facilities in order to run experiments in parallel. Hint: Bayesian optimization is provided as `syne_tune.optimizer.baselines.BayesianOptimization`.

    3. For this exercise, you need to run on an instance with at least 4 CPU cores. For one of the methods used above (random search, Bayesian optimization), run experiments with `n_workers=1`, `n_workers=2`, `n_workers=4`, and compare results (incumbent trajectories). At least for random search, you should observe linear scaling with respect to the number of workers. Hint: For robust results, you may have to average over several repetitions each.

2. *Advanced*. The goal of this exercise is to implement a new scheduler in Syne Tune.

    1. Create a virtual environment containing both the d2lbook[267] and syne-tune[268] sources.

    2. Implement the `LocalSearcher` from Exercise 2 in Section 19.2 as a new searcher in Syne Tune. Hint: Read this tutorial[269]. Alternatively, you may follow this example[270].

    3. Compare your new `LocalSearcher` with `RandomSearch` on the `DropoutMLP` benchmark.

Discussions[271].

## 19.4 Multi-Fidelity Hyperparameter Optimization

Training neural networks can be expensive even on moderate size datasets. Depending on the configuration space (Section 19.1.1), hyperparameter optimization requires tens to hundreds of function evaluations to find a well-performing hyperparameter configuration. As we have seen in Section 19.3, we can significantly speed up the overall wall-clock time of HPO by exploiting parallel resources, but this does not reduce the total amount of compute required.

In this section, we will show how the evaluation of hyperparameter configurations can be sped up. Methods such as random search allocate the same amount of resources (e.g., number of epochs, training data points) to each hyperparameter evaluation. Fig. 19.4.1 depicts learning curves of a set of neural networks trained with different hyperparameter configurations. After a few epochs we are already able to visually distinguish between well-performing and suboptimal configurations. However, the learning curves are noisy, and we might still require the full amount of 100 epochs to identify the best performing one.

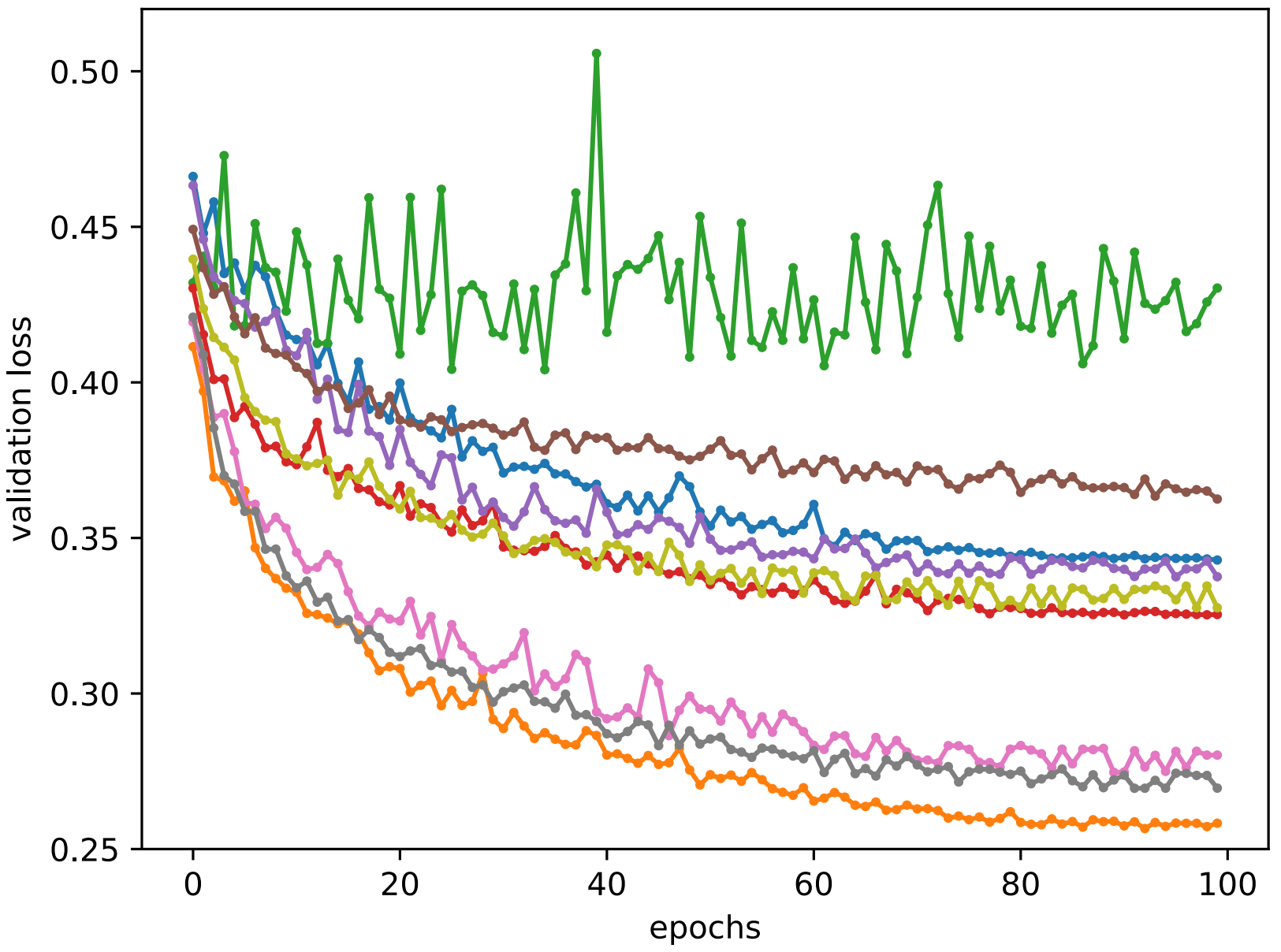

Fig. 19.4.1 Learning curves of random hyperparameter configurations

Multi-fidelity hyperparameter optimization allocates more resources to promising configurations and stop evaluations of poorly performing ones early. This speeds up the optimization process, since we can try a larger number of configurations for the same total amount of resources.

More formally, we expand our definition in Section 19.1.1, such that our objective function $f(\mathbf{x}, r)$ gets an additional input $r \in [r_{\mathrm{min}}, r_{max}]$, specifying the amount of resources that we are willing to spend for the evaluation of configuration $\mathbf{x}$. We assume that the error $f(\mathbf{x}, r)$ decreases with $r$, whereas the computational cost $c(\mathbf{x}, r)$ increases. Typically, $r$ represents the number of epochs for training the neural network, but it could also be the training subset size or the number of cross-validation folds.

```
from collections import defaultdict
import numpy as np
from scipy import stats
from d2l import torch as d2l

d2l.set_figsize()
```

### 19.4.1 Successive Halving

One of the simplest ways to adapt random search to the multi-fidelity setting is *successive halving* (Jamieson and Talwalkar, 2016, Karnin *et al.*, 2013). The basic idea is to start with $N$ configurations, for example randomly sampled from the configuration space, and to train each of them for $r_{\mathrm{min}}$ epochs only. We then discard a fraction of the worst performing trials and train the remaining ones for longer. Iterating this process, fewer trials run for longer, until at least one trial reaches $r_{max}$ epochs.

More formally, consider a minimum budget $r_{\mathrm{min}}$ (for example 1 epoch), a maximum budget $r_{max}$, for example `max_epochs` in our previous example, and a halving constant $\eta \in \{2, 3, \dots\}$. For simplicity, assume that $r_{max} = r_{\mathrm{min}}\eta^K$, with $K \in \mathbb{I}$ . The number of initial configurations is then $N = \eta^K$. Let us define the set of rungs $\mathcal{R} = \{r_{\mathrm{min}}, r_{\mathrm{min}}\eta, r_{\mathrm{min}}\eta^2, \dots, r_{max}\}$.

One round of successive halving proceeds as follows. We start with running $N$ trials until the first rung $r_{\mathrm{min}}$. Sorting the validation errors, we keep the top $1/\eta$ fraction (which amounts to $\eta^{K-1}$ configurations) and discard all the rest. The surviving trials are trained for the next rung ($r_{\mathrm{min}}\eta$ epochs), and the process is repeated. At each rung, a $1/\eta$ fraction of trials survives and their training continues with a $\eta$ times larger budget. With this particular choice of $N$, only a single trial will be trained to the full budget $r_{max}$. Once such a round of successive halving is done, we start the next one with a new set of initial configurations, iterating until the total budget is spent.

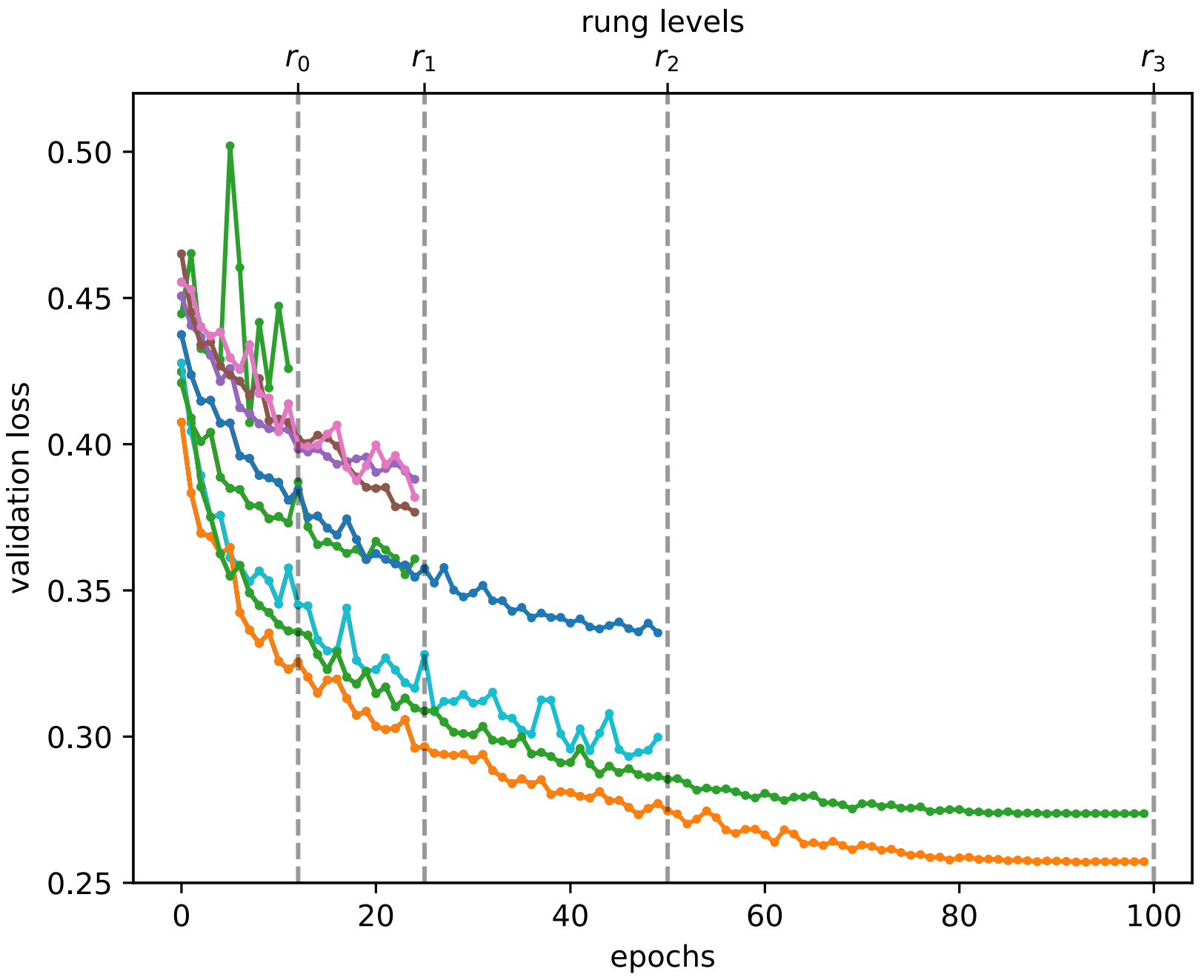

Fig. 19.4.2 Learning curves of random hyperparameter configurations.

We subclass the HPOScheduler base class from Section 19.2 in order to implement successive halving, allowing for a generic HPOSearcher object to sample configurations (which, in our example below, will be a RandomSearcher). Additionally, the user has to pass the minimum resource $r_{\mathrm{min}}$, the maximum resource $r_{max}$ and $\eta$ as input. Inside our scheduler, we maintain a queue of configurations that still need to be evaluated for the current rung $r_i$. We update the queue every time we jump to the next rung.

```
class SuccessiveHalvingScheduler(d2l.HPOScheduler):  #@save
    def __init__(self, searcher, eta, r_min, r_max, prefact=1):
        self.save_hyperparameters()
        # Compute K, which is later used to determine the number of␣
↪configurations
        self.K = int(np.log(r_max / r_min) / np.log(eta))
        # Define the rungs
        self.rung_levels = [r_min * eta ** k for k in range(self.K + 1)]
        if r_max not in self.rung_levels:
            # The final rung should be r_max
            self.rung_levels.append(r_max)
            self.K += 1
        # Bookkeeping
        self.observed_error_at_rungs = defaultdict(list)
        self.all_observed_error_at_rungs = defaultdict(list)
        # Our processing queue
        self.queue = []
```

In the beginning our queue is empty, and we fill it with $n = \mathrm{prefact} \cdot \eta^K$ configurations, which are first evaluated on the smallest rung $r_{\mathrm{min}}$. Here, prefact allows us to reuse our code in a different context. For the purpose of this section, we fix prefact $= 1$. Every time resources become available and the HPOTuner object queries the suggest function, we return an element from the queue. Once we finish one round of successive halving, which means that we evaluated all surviving configurations on the highest resource level $r_{max}$ and our queue is empty, we start the entire process again with a new, randomly sampled set of configurations.

```
@d2l.add_to_class(SuccessiveHalvingScheduler)  #@save
def suggest(self):
    if len(self.queue) == 0:
        # Start a new round of successive halving
        # Number of configurations for the first rung:
        n0 = int(self.prefact * self.eta ** self.K)
        for _ in range(n0):
            config = self.searcher.sample_configuration()
            config["max_epochs"] = self.r_min  # Set r = r_min
            self.queue.append(config)
    # Return an element from the queue
    return self.queue.pop()
```

When we collected a new data point, we first update the searcher module. Afterwards we check if we already collect all data points on the current rung. If so, we sort all configurations and push the top $\frac{1}{\eta}$ configurations into the queue.

```
@d2l.add_to_class(SuccessiveHalvingScheduler)  #@save
def update(self, config: dict, error: float, info=None):
    ri = int(config["max_epochs"])  # Rung r_i
    # Update our searcher, e.g if we use Bayesian optimization later
    self.searcher.update(config, error, additional_info=info)
    self.all_observed_error_at_rungs[ri].append((config, error))
    if ri < self.r_max:
        # Bookkeeping
        self.observed_error_at_rungs[ri].append((config, error))
        # Determine how many configurations should be evaluated on this rung
        ki = self.K - self.rung_levels.index(ri)
        ni = int(self.prefact * self.eta ** ki)
        # If we observed all configuration on this rung r_i, we estimate the
        # top 1 / eta configuration, add them to queue and promote them for
        # the next rung r_{i+1}
        if len(self.observed_error_at_rungs[ri]) >= ni:
            kiplus1 = ki - 1
            niplus1 = int(self.prefact * self.eta ** kiplus1)
            best_performing_configurations = self.get_top_n_configurations(
                rung_level=ri, n=niplus1
            )
            riplus1 = self.rung_levels[self.K - kiplus1]  # r_{i+1}
            # Queue may not be empty: insert new entries at the beginning
            self.queue = [
                dict(config, max_epochs=riplus1)
                for config in best_performing_configurations
            ] + self.queue
            self.observed_error_at_rungs[ri] = []  # Reset
```

Configurations are sorted based on their observed performance on the current rung.

```
@d2l.add_to_class(SuccessiveHalvingScheduler)  #@save
def get_top_n_configurations(self, rung_level, n):
    rung = self.observed_error_at_rungs[rung_level]
    if not rung:
        return []
    sorted_rung = sorted(rung, key=lambda x: x[1])
    return [x[0] for x in sorted_rung[:n]]
```

Let us see how successive halving is doing on our neural network example. We will use $r_{\min} = 2$, $\eta = 2$, $r_{max} = 10$, so that rung levels are $2, 4, 8, 10$.

```
min_number_of_epochs = 2
max_number_of_epochs = 10
eta = 2
num_gpus=1

config_space = {
    "learning_rate": stats.loguniform(1e-2, 1),
    "batch_size": stats.randint(32, 256),
}
initial_config = {
    "learning_rate": 0.1,
```

```
    "batch_size": 128,
}
```

We just replace the scheduler with our new `SuccessiveHalvingScheduler`.

```
searcher = d2l.RandomSearcher(config_space, initial_config=initial_config)
scheduler = SuccessiveHalvingScheduler(
    searcher=searcher,
    eta=eta,
    r_min=min_number_of_epochs,
    r_max=max_number_of_epochs,
)
tuner = d2l.HPOTuner(
    scheduler=scheduler,
    objective=d2l.hpo_objective_lenet,
)
tuner.run(number_of_trials=30)
```

```
    error = 0.2517373561859131, runtime = 56.05348062515259
```

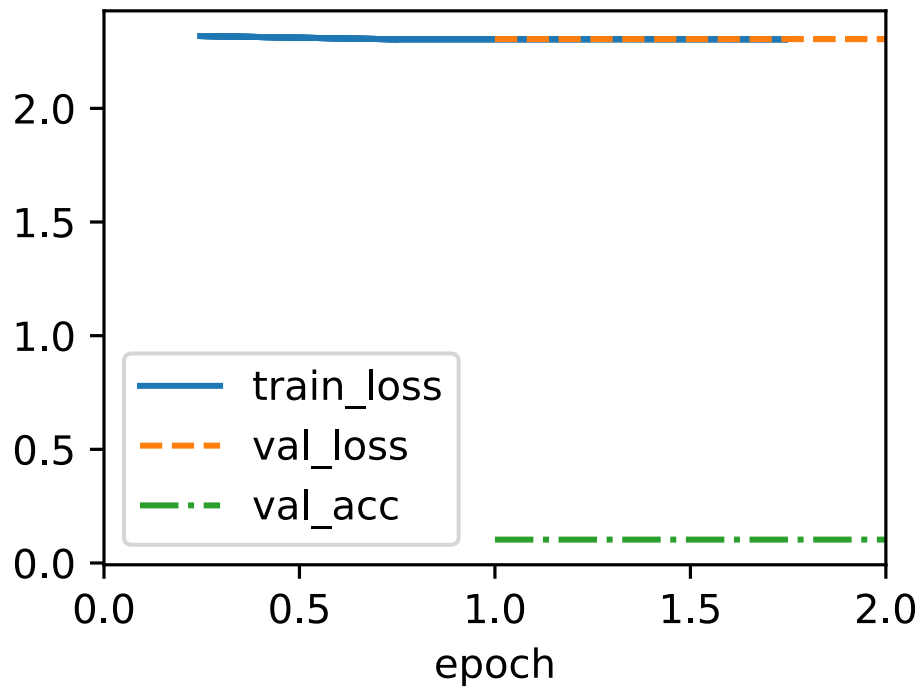

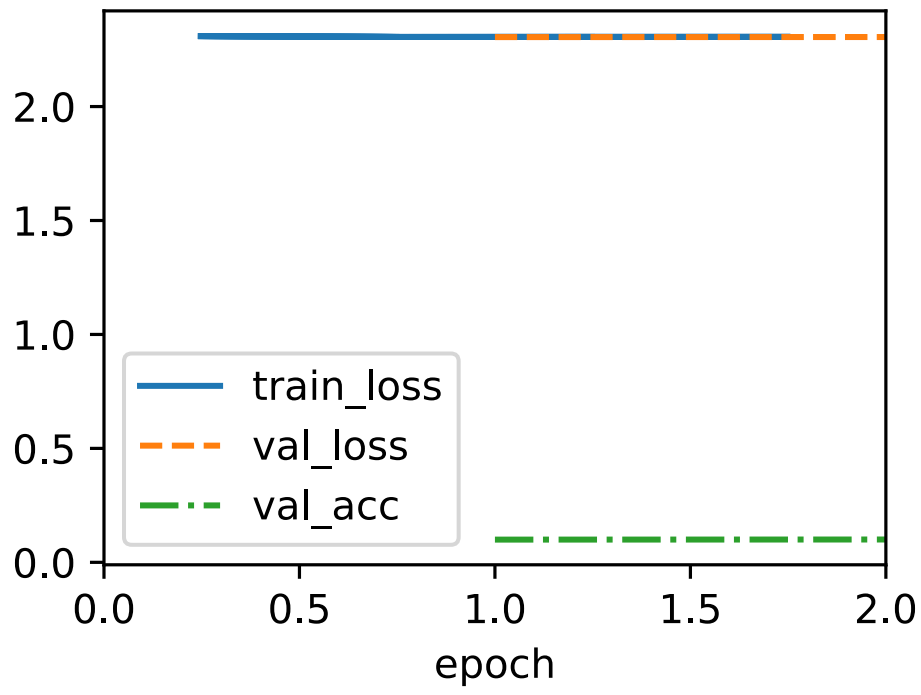

We can visualize the learning curves of all configurations that we evaluated. Most of the configurations are stopped early and only the better performing configurations survive until

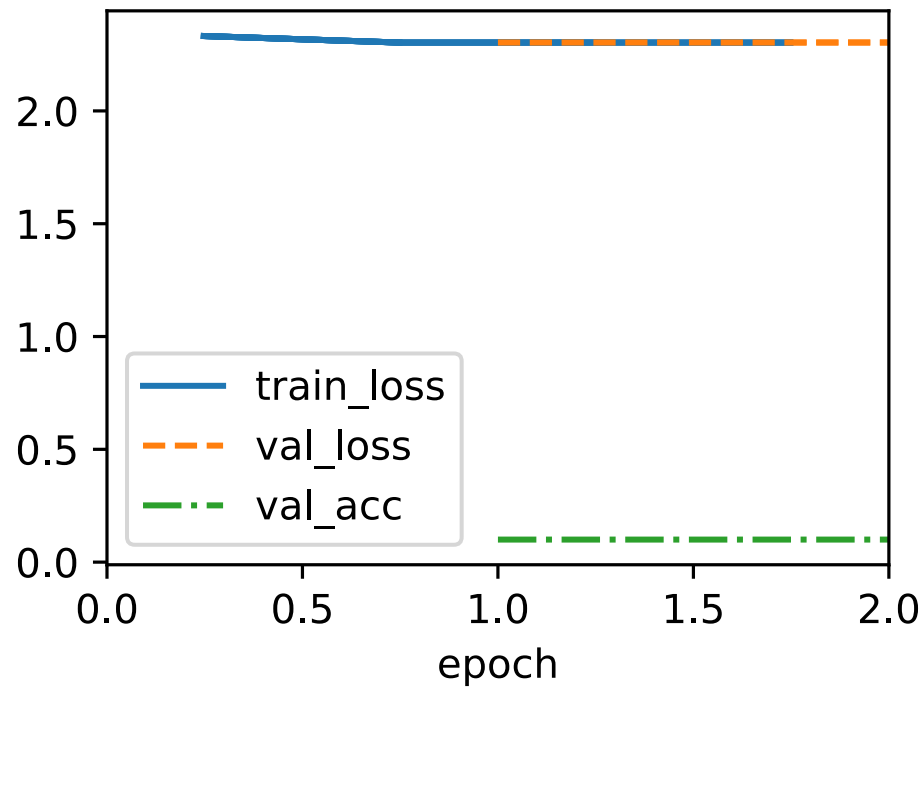

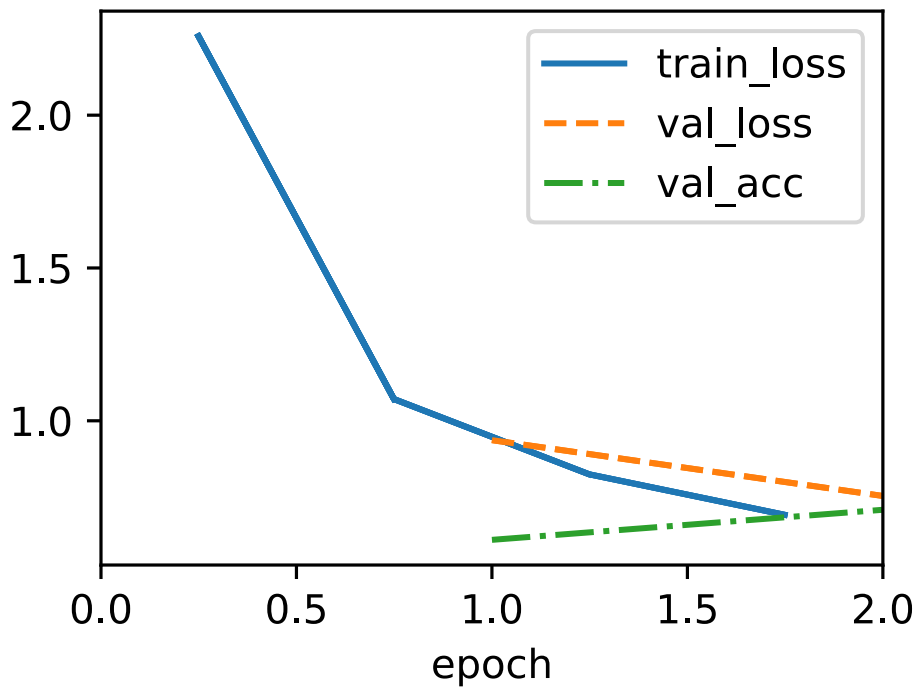

$r_{max}$. Compare this to vanilla random search, which would allocate $r_{max}$ to every configuration.

```
for rung_index, rung in scheduler.all_observed_error_at_rungs.items():
    errors = [xi[1] for xi in rung]
    d2l.plt.scatter([rung_index] * len(errors), errors)
d2l.plt.xlim(min_number_of_epochs - 0.5, max_number_of_epochs + 0.5)
d2l.plt.xticks(
    np.arange(min_number_of_epochs, max_number_of_epochs + 1),
```

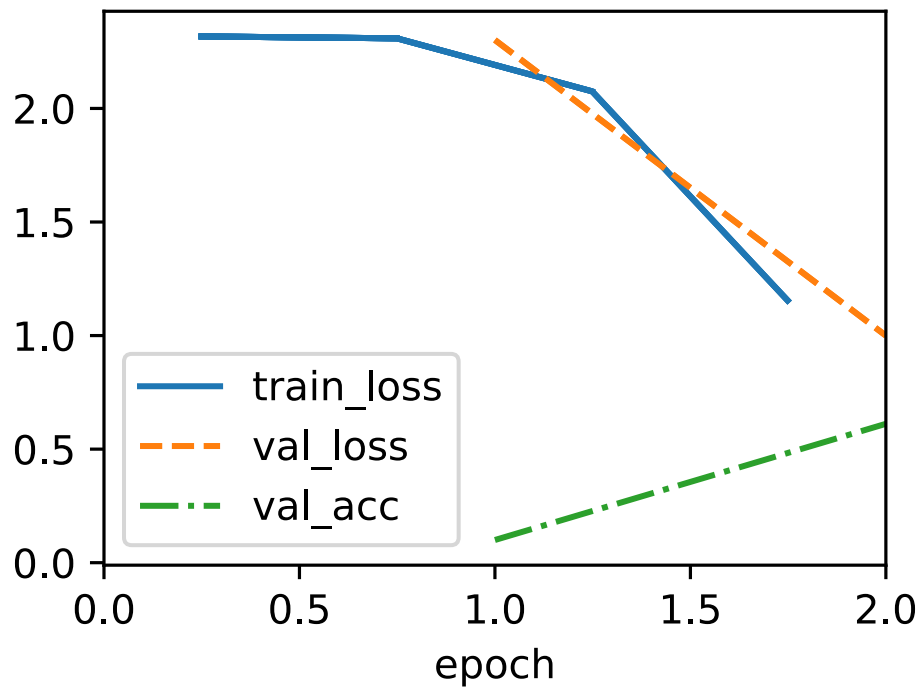

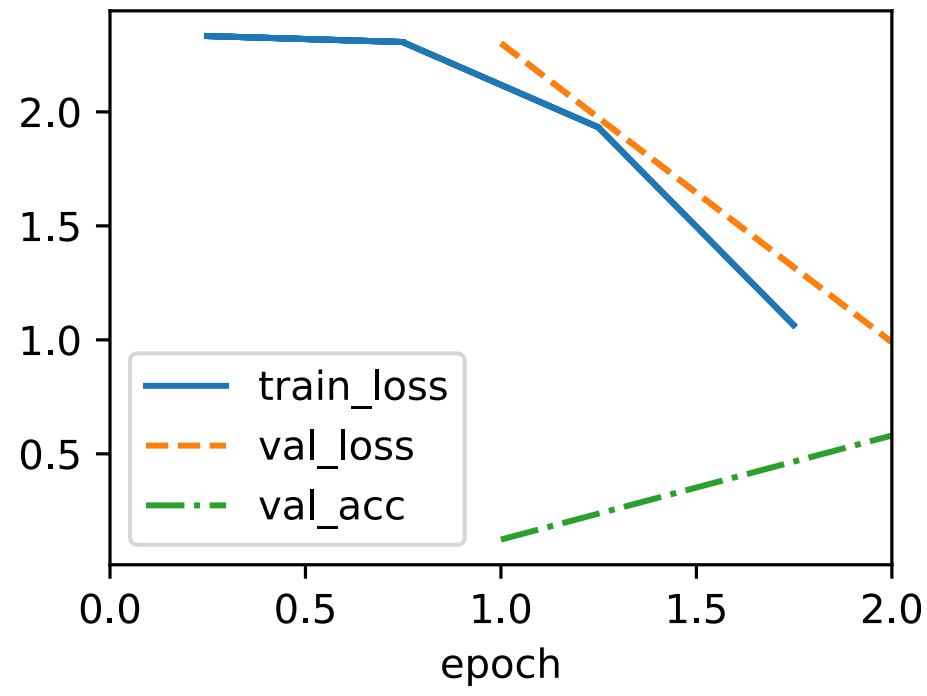

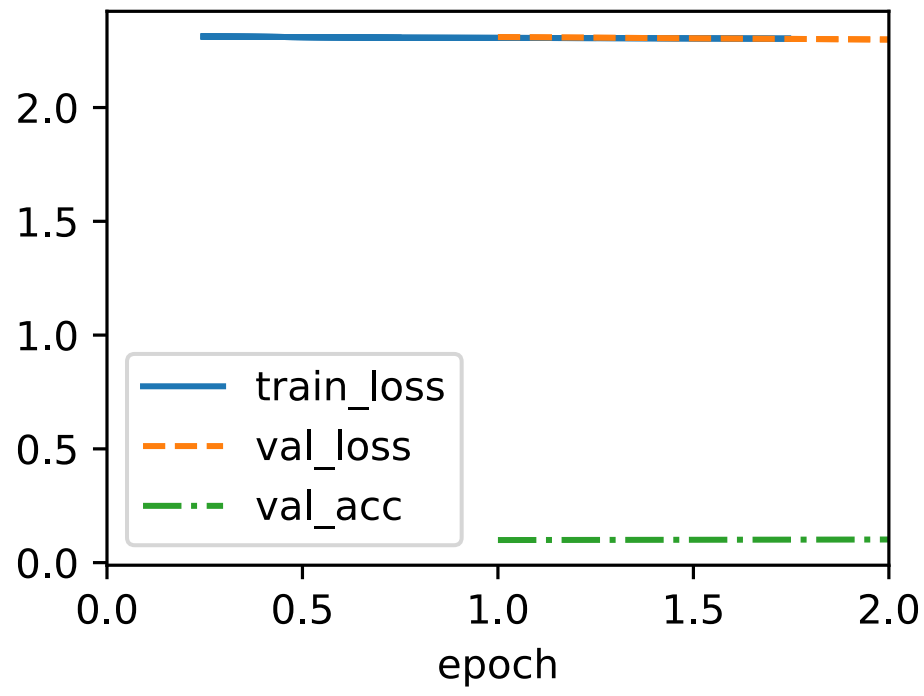

```
    np.arange(min_number_of_epochs, max_number_of_epochs + 1)
)
d2l.plt.ylabel("validation error")
d2l.plt.xlabel("epochs")
```

```
Text(0.5, 0, 'epochs')
```

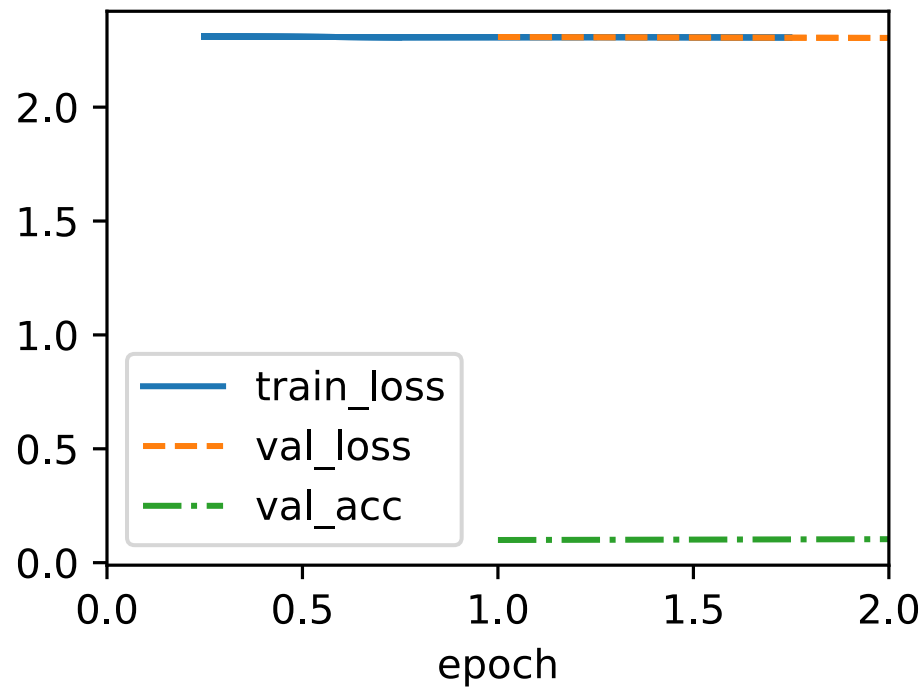

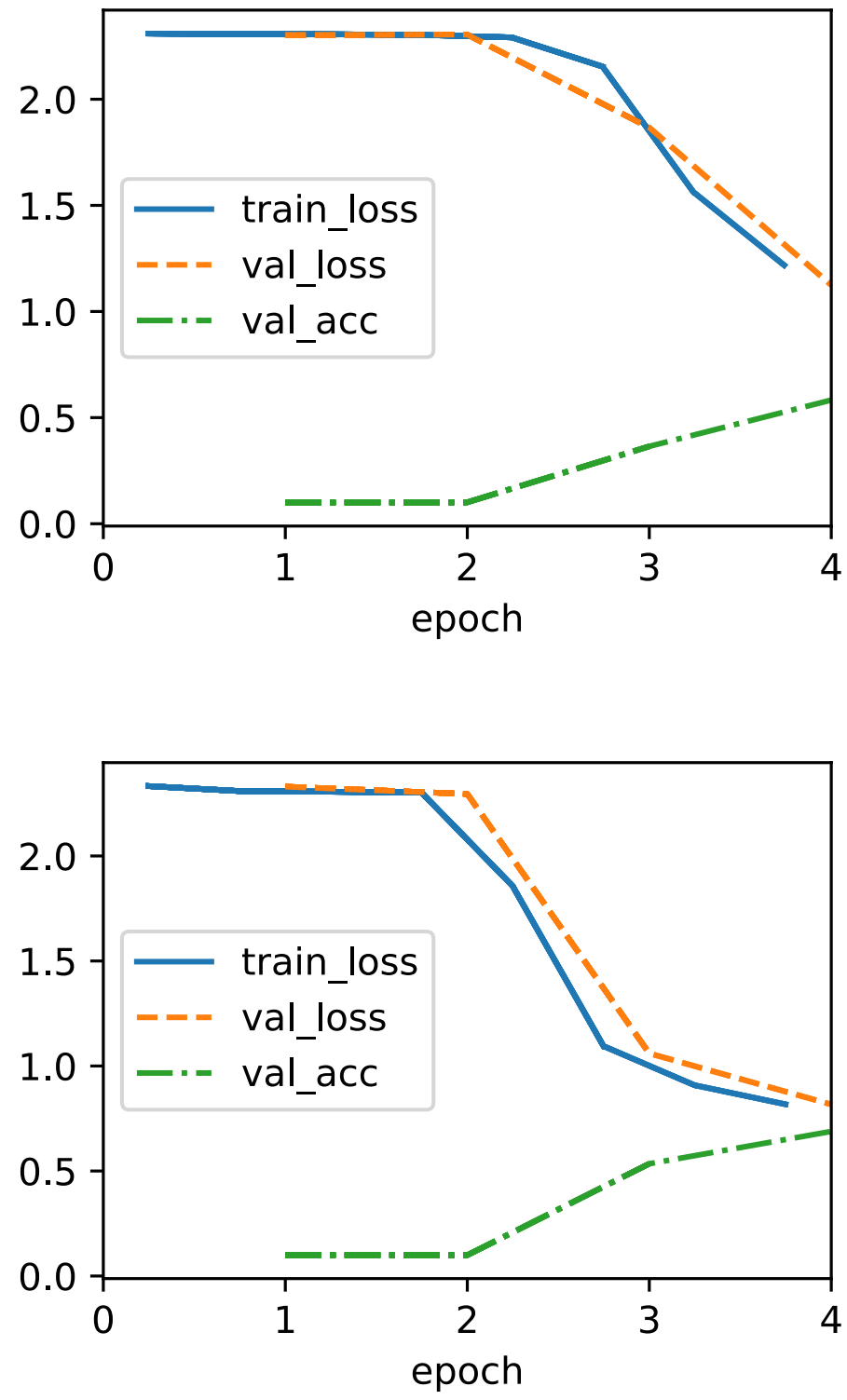

Finally, note some slight complexity in our implementation of `SuccessiveHalvingScheduler`. Say that a worker is free to run a job, and `suggest` is called when the current rung has almost been completely filled, but another worker is still busy with an evaluation. Since we lack the metric value from this worker, we cannot determine the top $1/\eta$ fraction to open up the next rung. On the other hand, we want to assign a job to our free worker, so it does not remain idle. Our solution is to start a new round of successive halving and assign our worker to the first trial there. However, once a rung is completed in `update`, we make sure to insert new configurations at the beginning of the queue, so they take precedence over configurations from the next round.

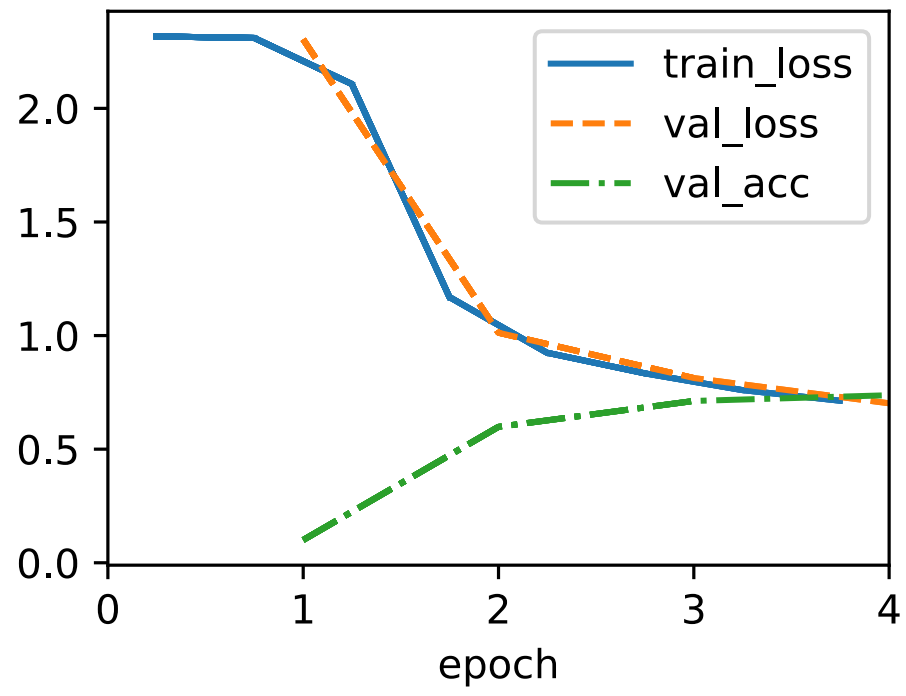

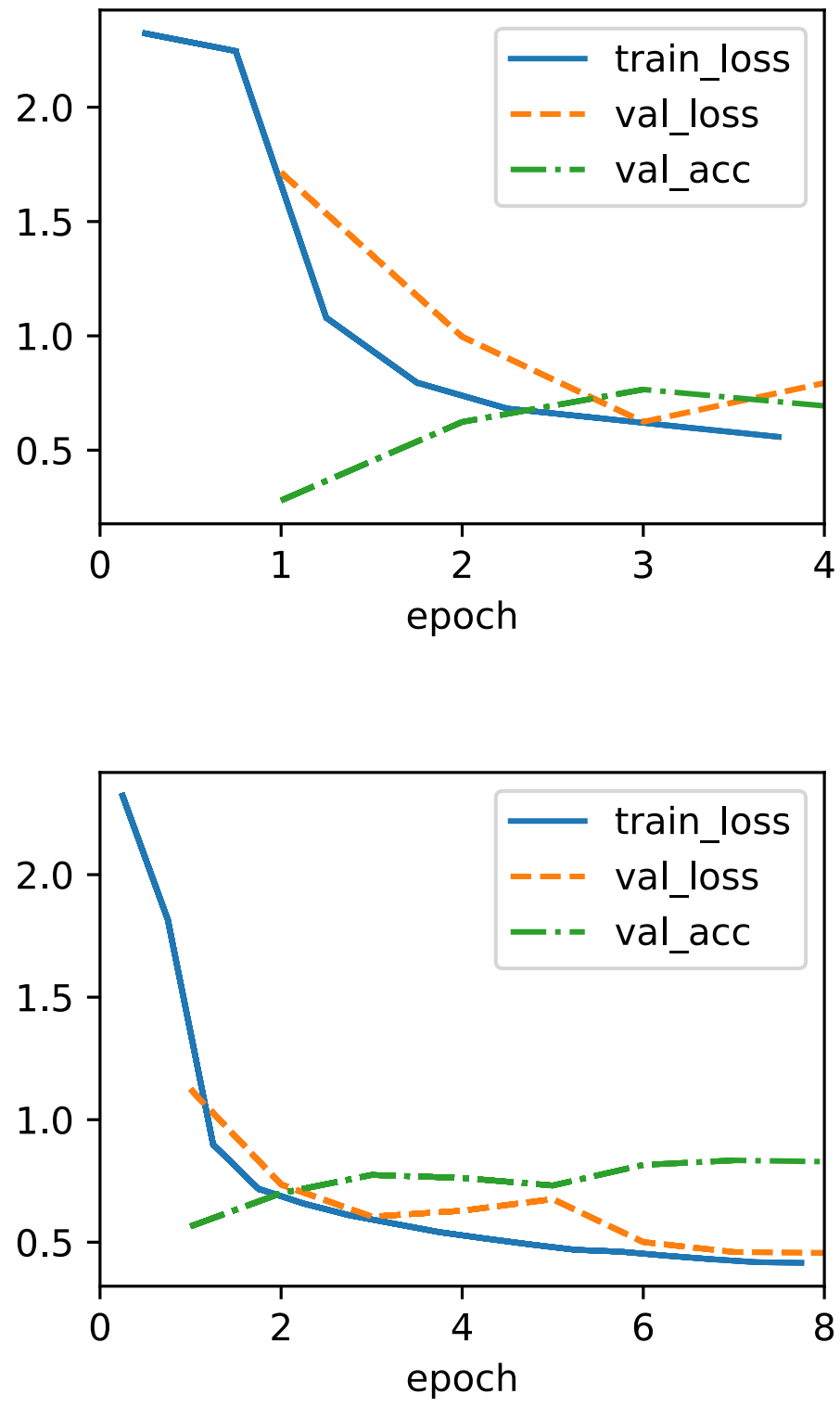

## 19.4.2 Summary

In this section, we introduced the concept of multi-fidelity hyperparameter optimization, where we assume to have access to cheap-to-evaluate approximations of the objective function, such as validation error after a certain number of epochs of training as proxy to validation error after the full number of epochs. Multi-fidelity hyperparameter optimization allows to reduce the overall computation of the HPO instead of just reducing the wall-clock time.

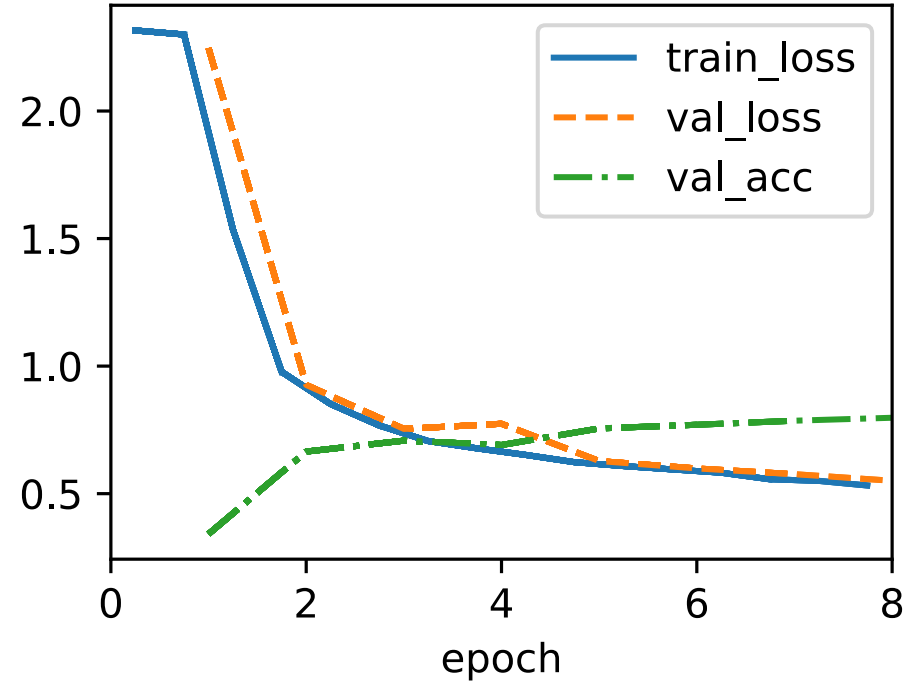

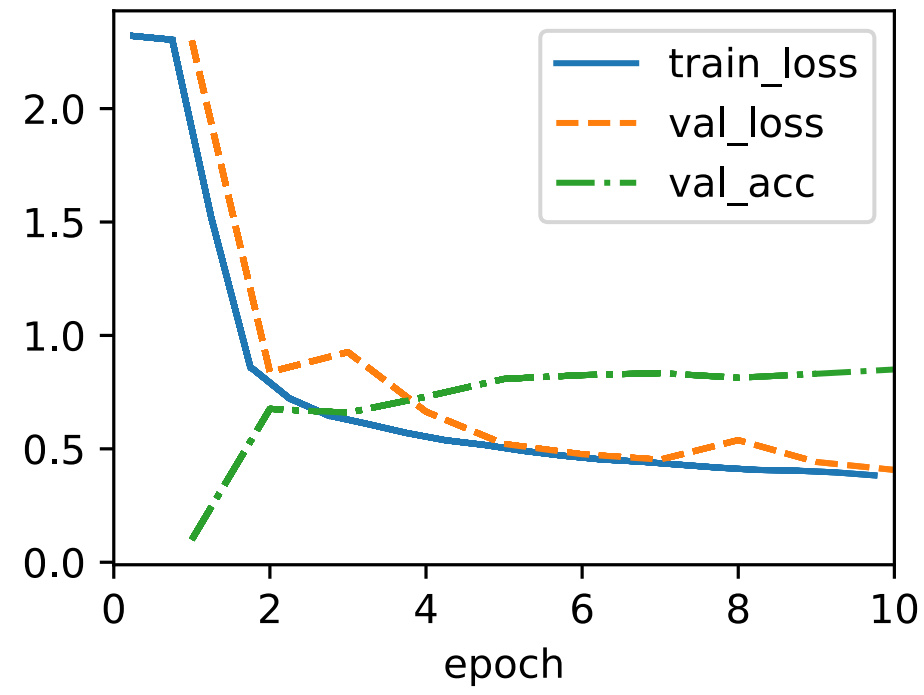

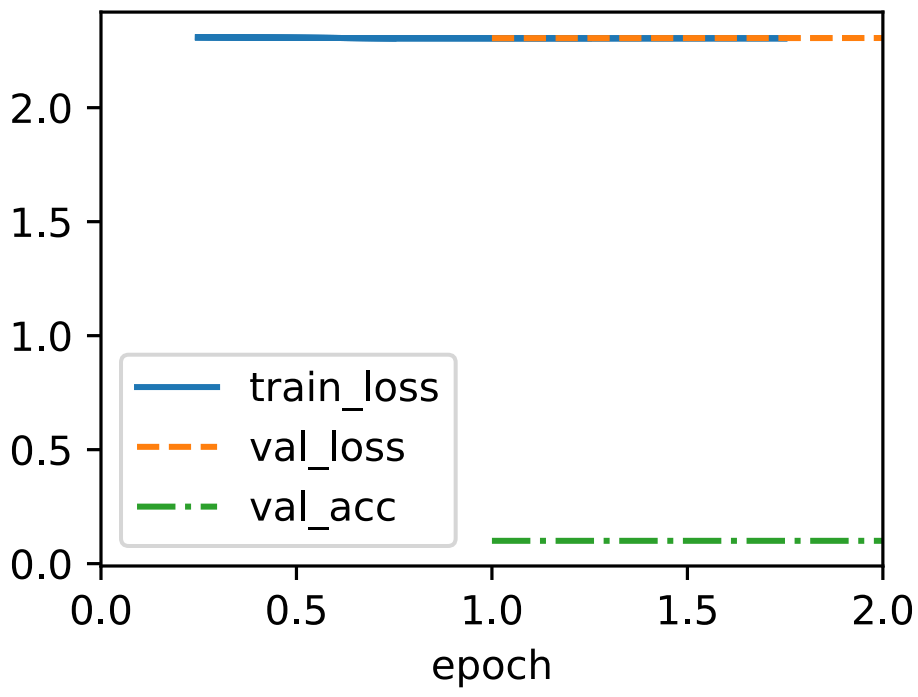

We implemented and evaluated successive halving, a simple yet efficient multi-fidelity HPO algorithm.

272

Discussions[272].

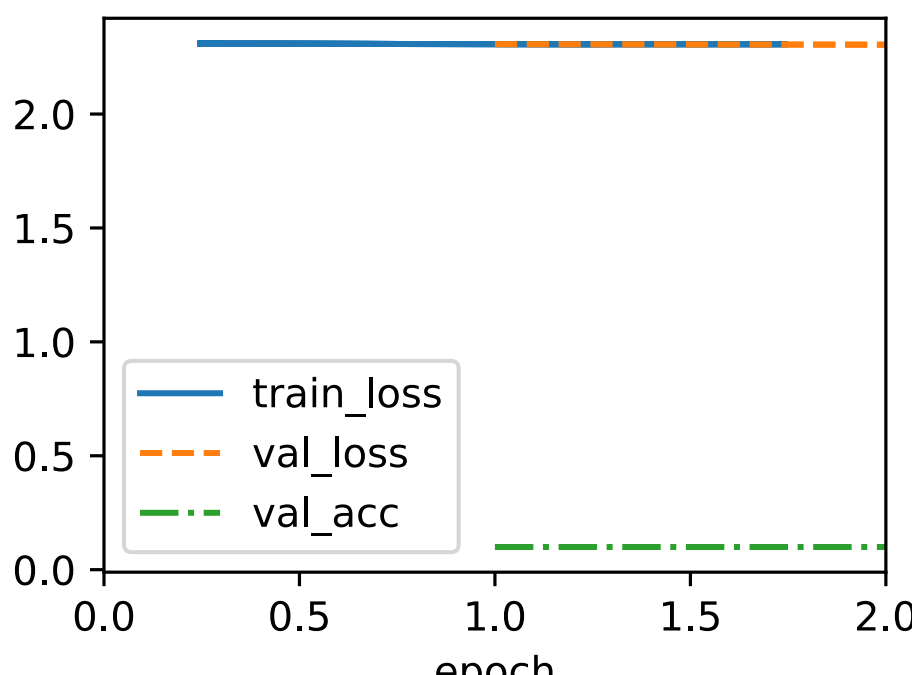

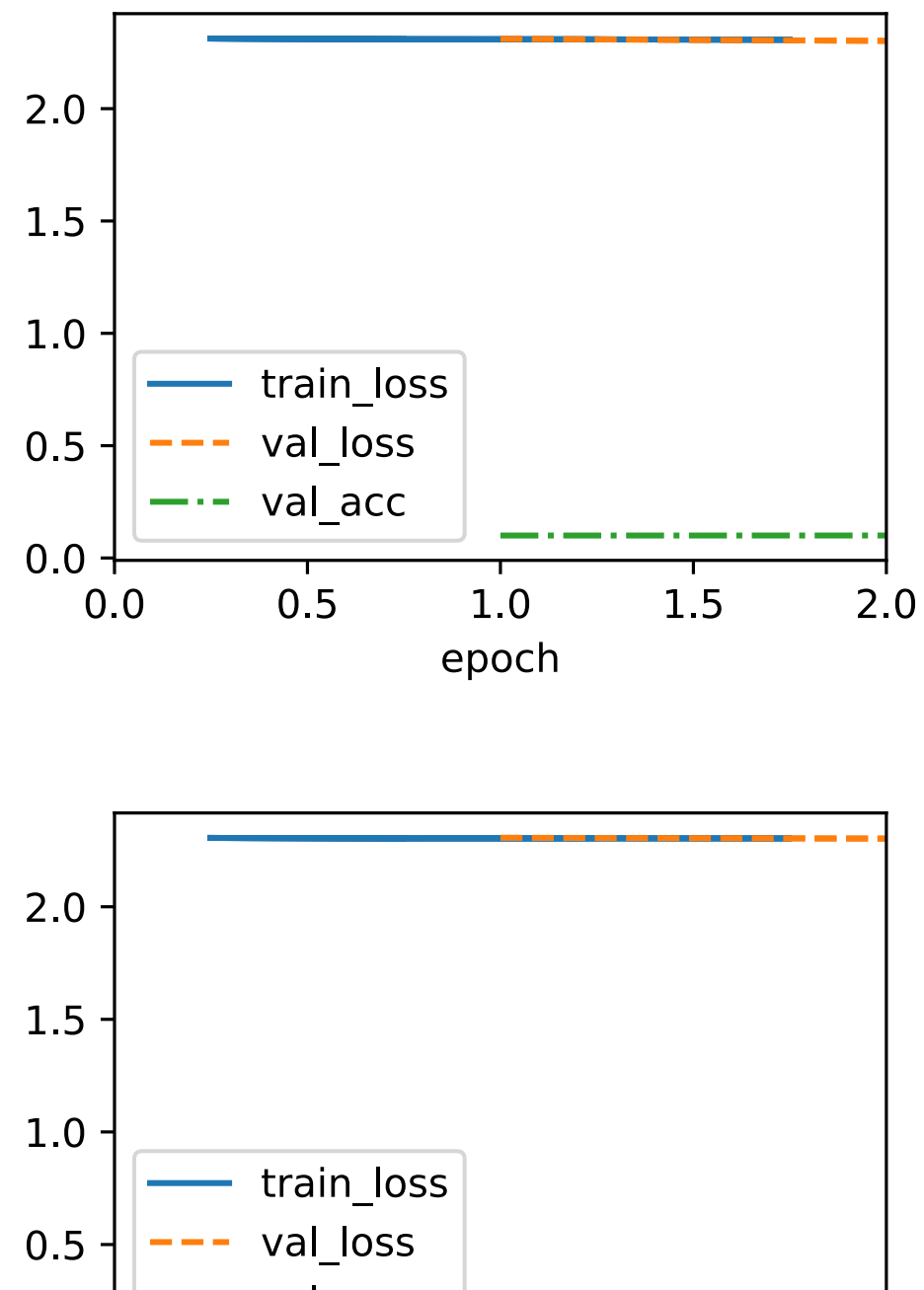

## 19.5 Asynchronous Successive Halving

As we have seen in Section 19.3, we can accelerate HPO by distributing the evaluation of hyperparameter configurations across either multiple instances or multiples CPUs / GPUs on a single instance. However, compared to random search, it is not straightforward to run successive halving (SH) asynchronously in a distributed setting. Before we can decide which configuration to run next, we first have to collect all observations at the current rung level. This requires to synchronize workers at each rung level. For example, for the lowest rung level $r_{\mathrm{min}}$, we first have to evaluate all $N = \eta^K$ configurations, before we can promote the $\frac{1}{\eta}$ of them to the next rung level.

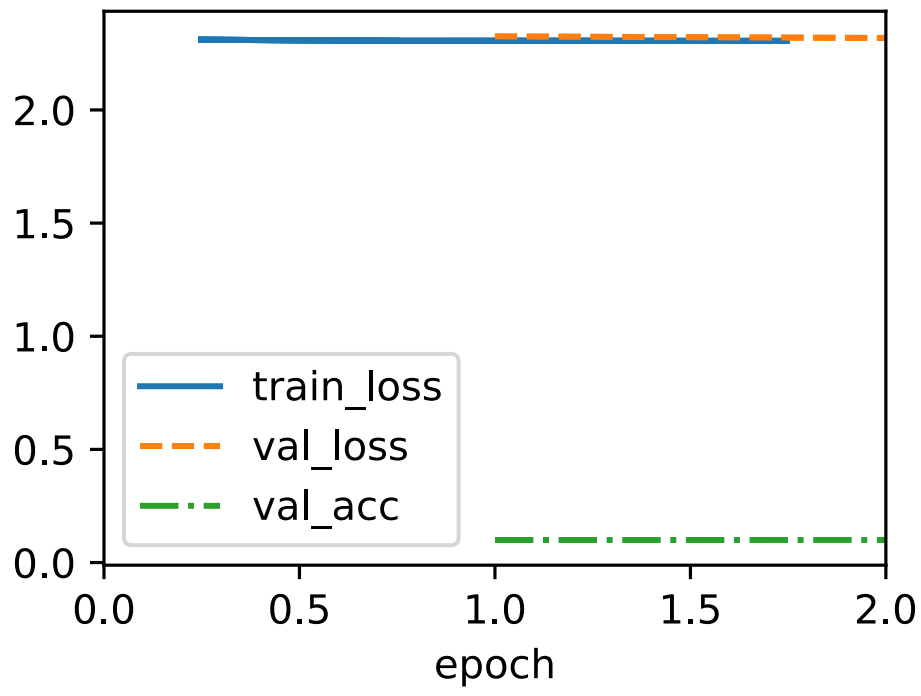

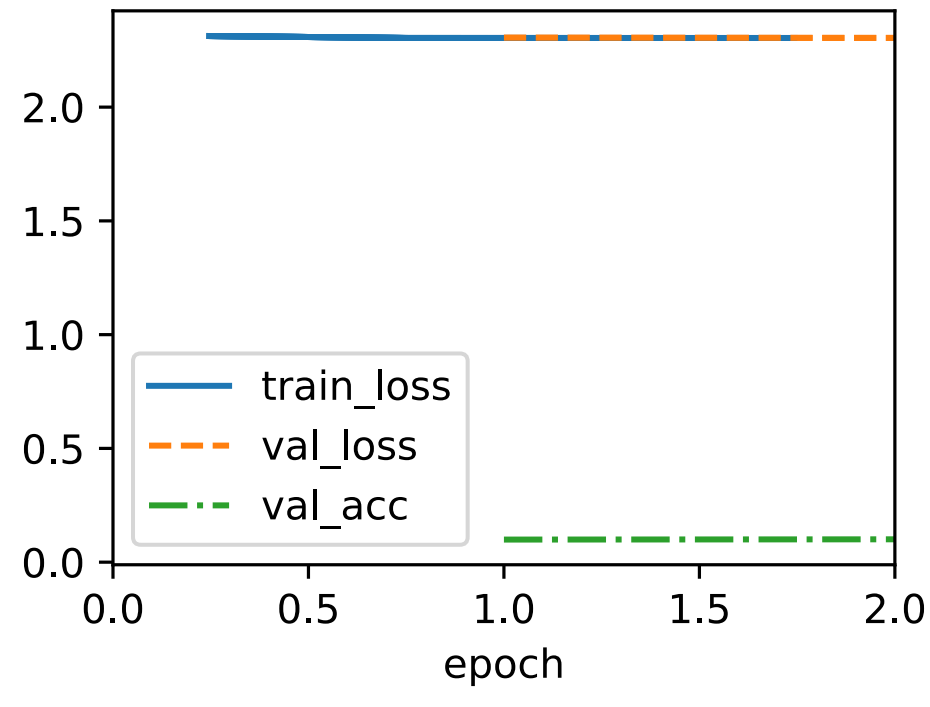

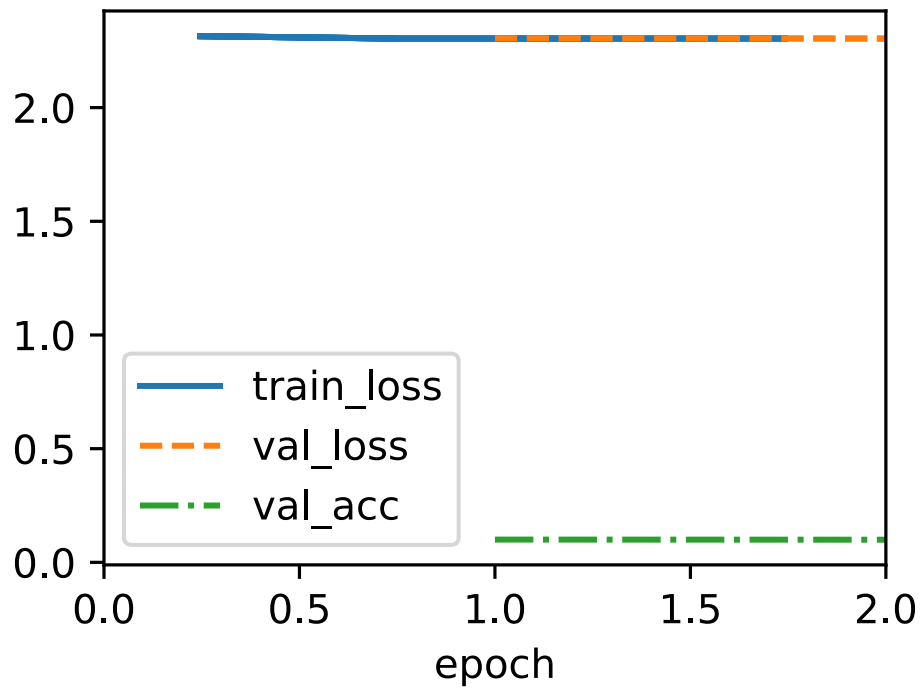

In any distributed system, synchronization typically implies idle time for workers. First, we often observe high variations in training time across hyperparameter configurations. For example, assuming the number of filters per layer is a hyperparameter, then networks with less filters finish training faster than networks with more filters, which implies idle worker time due to stragglers. Moreover, the number of slots in a rung level is not always a multiple of the number of workers, in which case some workers may even sit idle for a full batch.

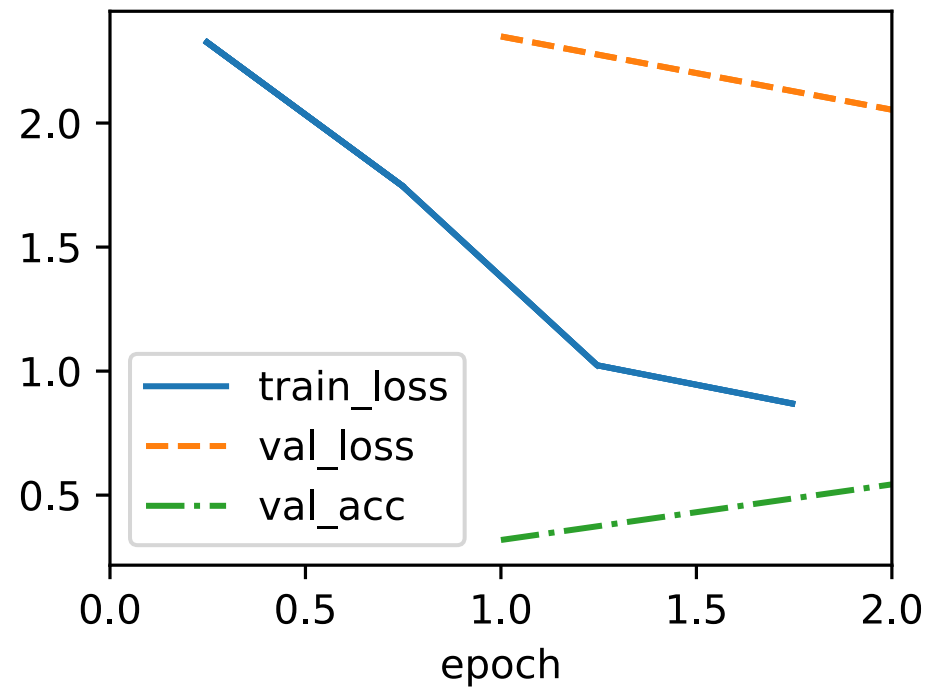

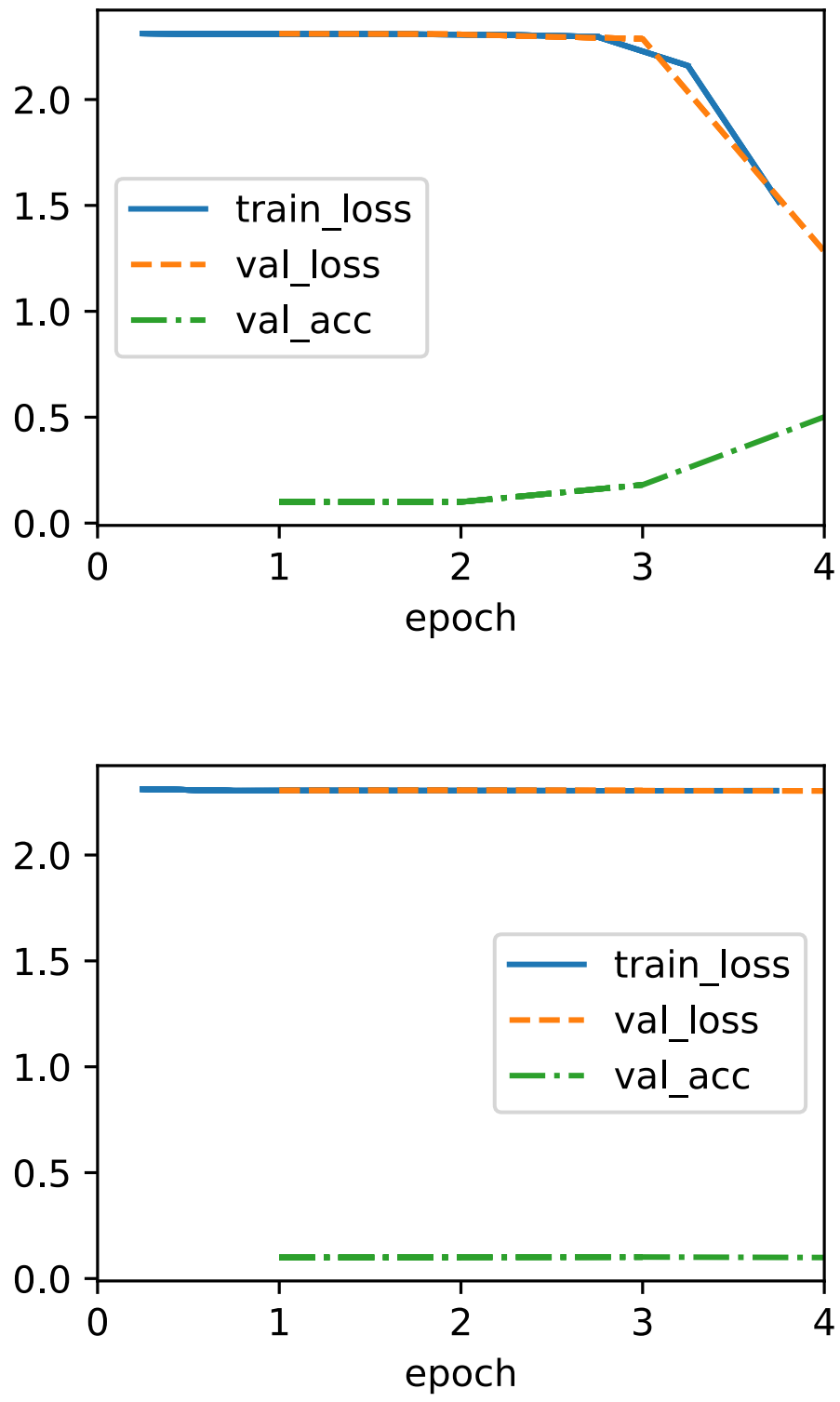

Figure Fig. 19.5.1 shows the scheduling of synchronous SH with $\eta = 2$ for four different trials with two workers. We start with evaluating Trial-0 and Trial-1 for one epoch and immediately continue with the next two trials once they are finished. We first have to wait until Trial-2 finishes, which takes substantially more time than the other trials, before we can promote the best two trials, i.e., Trial-0 and Trial-3 to the next rung level. This causes idle time for Worker-1. Then, we continue with Rung 1. Also, here Trial-3 takes longer than Trial-0, which

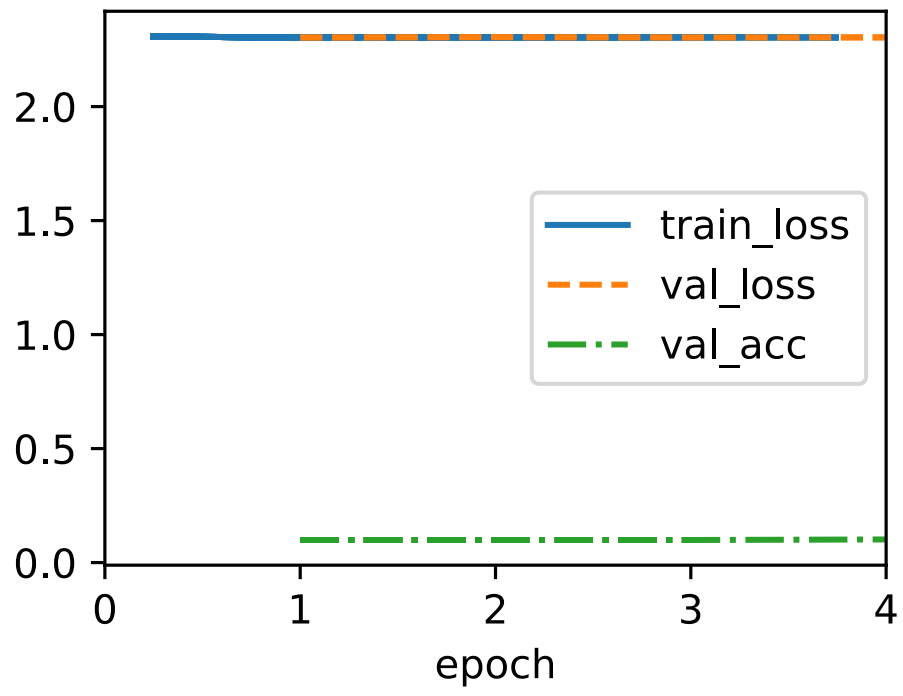

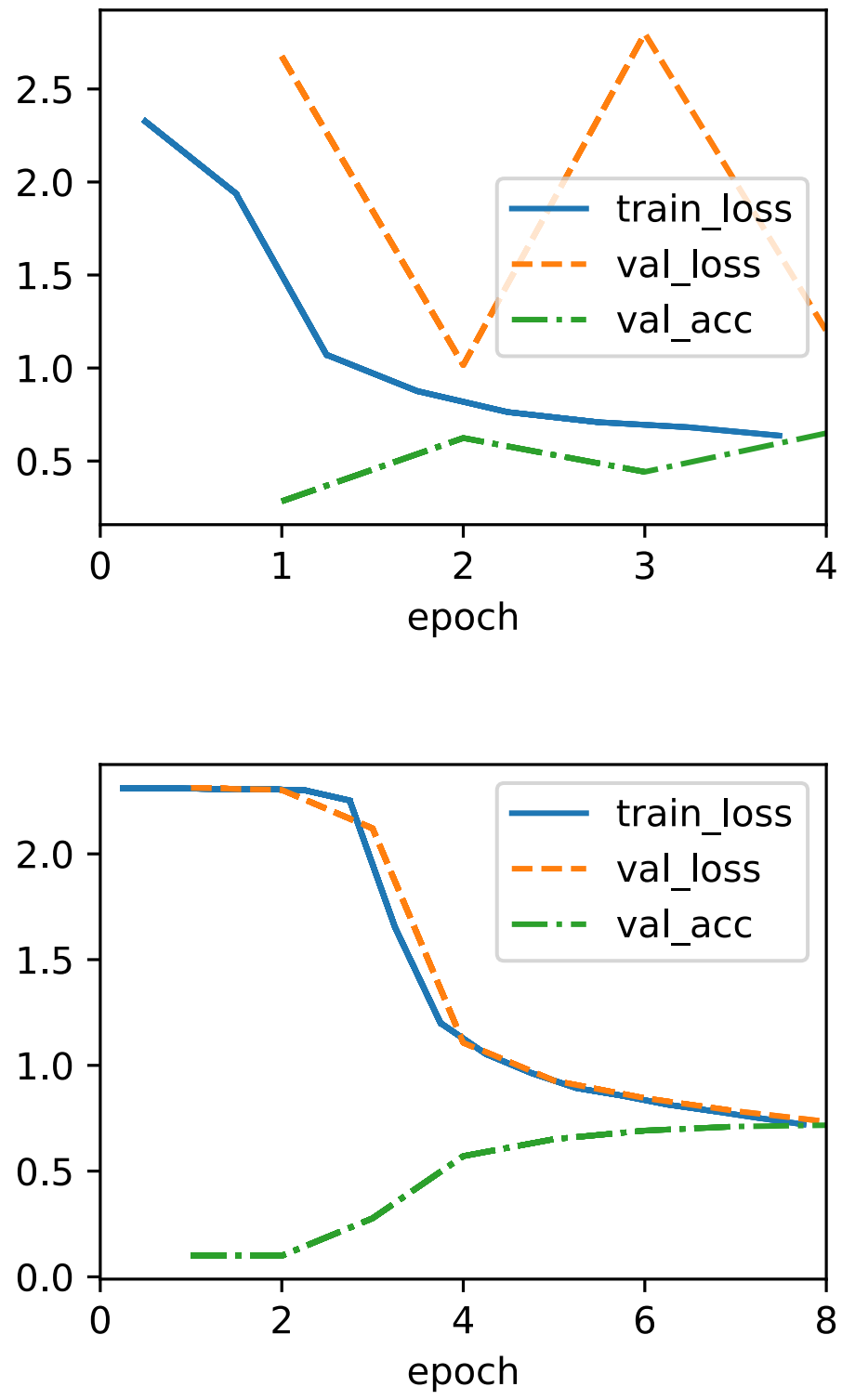

leads to an additional ideling time of Worker-0. Once, we reach Rung-2, only the best trial, Trial-0, remains which occupies only one worker. To avoid that Worker-1 idles during that time, most implementaitons of SH continue already with the next round, and start evaluating new trials (e.g Trial-4) on the first rung.

Asynchronous successive halving (ASHA) (Li *et al.*, 2018) adapts SH to the asynchronous parallel scenario. The main idea of ASHA is to promote configurations to the next rung level as soon as we collected at least $\eta$ observations on the current rung level. This decision rule may lead to suboptimal promotions: configurations can be promoted to the next rung level, which in hindsight do not compare favourably against most others at the same rung level. On the other hand, we get rid of all synchronization points this way. In practice, such suboptimal initial promotions have only a modest impact on performance, not only because the ranking of hyperparameter configurations is often fairly consistent across rung levels, but also because rungs grow over time and reflect the distribution of metric values at this level better and better. If a worker is free, but no configuration can be promoted, we start a new configuration with $r = r_{\min}$, i.e the first rung level.

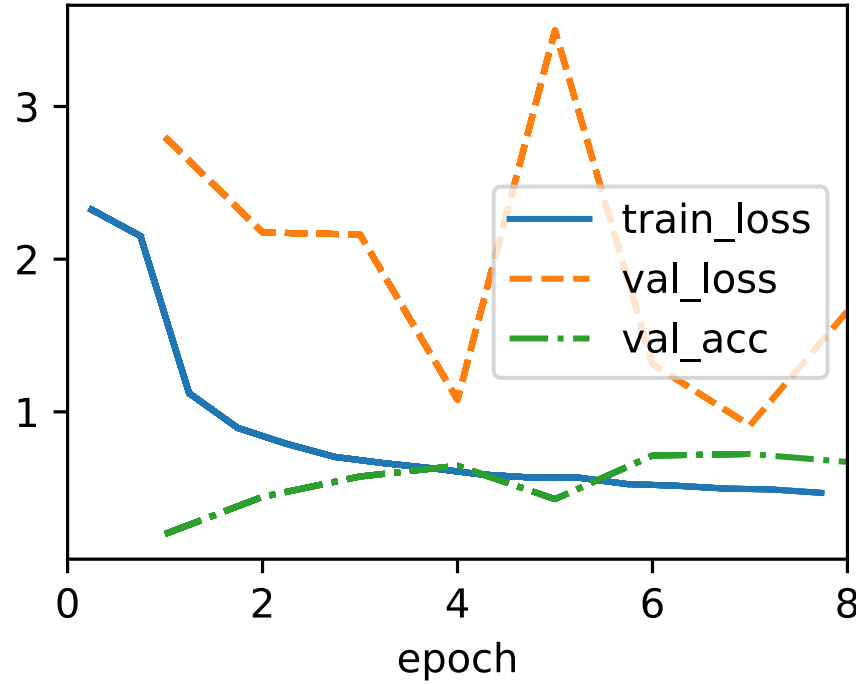

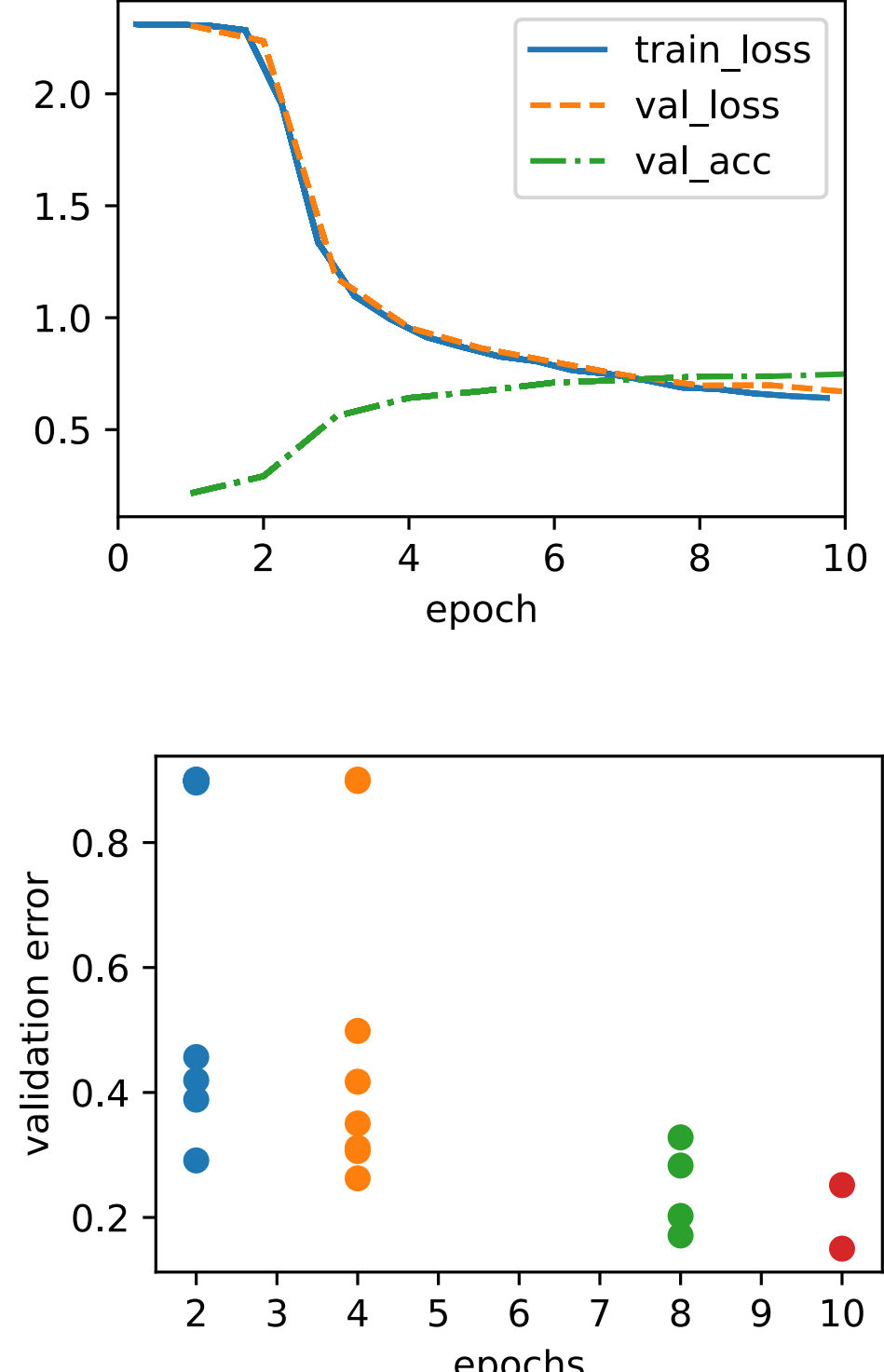

Fig. 19.5.2 shows the scheduling of the same configurations for ASHA. Once Trial-1 finishes,

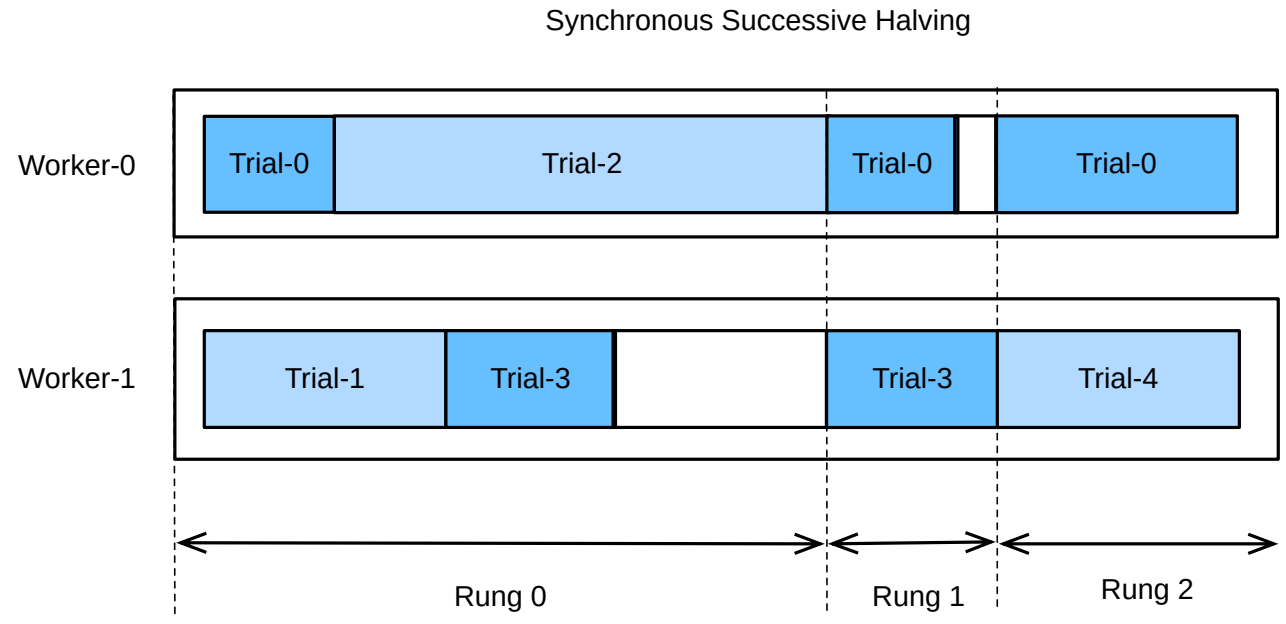

Fig. 19.5.1 Synchronous successive halving with two workers.

we collect the results of two trials (i.e Trial-0 and Trial-1) and immediately promote the better of them (Trial-0) to the next rung level. After Trial-0 finishes on rung 1, there are too few trials there in order to support a further promotion. Hence, we continue with rung 0 and evaluate Trial-3. Once Trial-3 finishes, Trial-2 is still pending. At this point we have 3 trials evaluated on rung 0 and one trial evaluated already on rung 1. Since Trial-3 performs worse than Trial-0 at rung 0, and $\eta = 2$, we cannot promote any new trial yet, and Worker-1 starts Trial-4 from scratch instead. However, once Trial-2 finishes and scores worse than Trial-3, the latter is promoted towards rung 1. Afterwards, we collected 2 evaluations on rung 1, which means we can now promote Trial-0 towards rung 2. At the same time, Worker-1 continues with evaluating new trials (i.e., Trial-5) on rung 0.

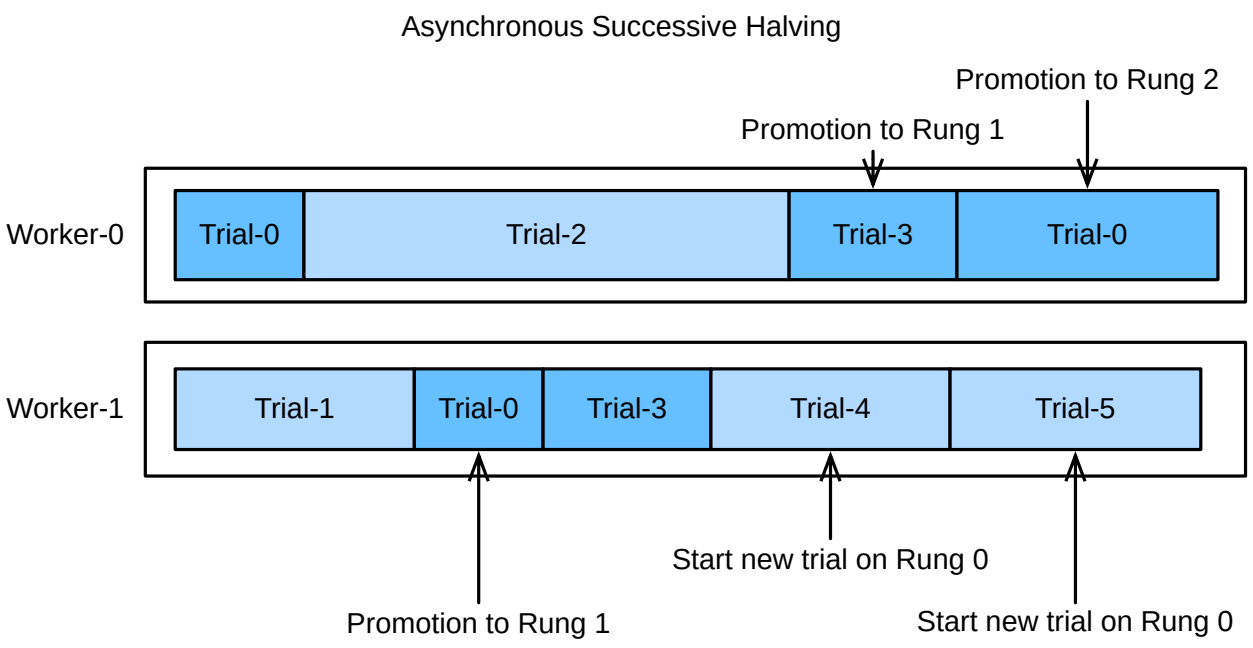

Fig. 19.5.2 Asynchronous successive halving (ASHA) with two workers.

```
import logging
from d2l import torch as d2l

logging.basicConfig(level=logging.INFO)
import matplotlib.pyplot as plt
from syne_tune import StoppingCriterion, Tuner
from syne_tune.backend.python_backend import PythonBackend
from syne_tune.config_space import loguniform, randint
from syne_tune.experiments import load_experiment
from syne_tune.optimizer.baselines import ASHA
```

```
INFO:root:SageMakerBackend is not imported since dependencies are missing. You␣
↪can install them with
   pip install 'syne-tune[extra]'
AWS dependencies are not imported since dependencies are missing. You can␣
↪install them with
   pip install 'syne-tune[aws]'
or (for everything)
   pip install 'syne-tune[extra]'
AWS dependencies are not imported since dependencies are missing. You can␣
↪install them with
   pip install 'syne-tune[aws]'
or (for everything)
   pip install 'syne-tune[extra]'
INFO:root:Ray Tune schedulers and searchers are not imported since␣
↪dependencies are missing. You can install them with
```

(continued from previous page)

```
   pip install 'syne-tune[raytune]'
or (for everything)
   pip install 'syne-tune[extra]'
```

## 19.5.1 Objective Function

We will use *Syne Tune* with the same objective function as in Section 19.3.

```
def hpo_objective_lenet_synetune(learning_rate, batch_size, max_epochs):
    from syne_tune import Reporter
    from d2l import torch as d2l

    model = d2l.LeNet(lr=learning_rate, num_classes=10)
    trainer = d2l.HPOTrainer(max_epochs=1, num_gpus=1)
    data = d2l.FashionMNIST(batch_size=batch_size)
    model.apply_init([next(iter(data.get_dataloader(True)))[0]], d2l.init_cnn)
    report = Reporter()
    for epoch in range(1, max_epochs + 1):
        if epoch == 1:
            # Initialize the state of Trainer
            trainer.fit(model=model, data=data)
        else:
            trainer.fit_epoch()
        validation_error = trainer.validation_error().cpu().detach().numpy()
        report(epoch=epoch, validation_error=float(validation_error))
```

We will also use the same configuration space as before:

```
min_number_of_epochs = 2
max_number_of_epochs = 10
eta = 2

config_space = {
    "learning_rate": loguniform(1e-2, 1),
    "batch_size": randint(32, 256),
    "max_epochs": max_number_of_epochs,
}
initial_config = {
    "learning_rate": 0.1,
    "batch_size": 128,
}
```

## 19.5.2 Asynchronous Scheduler

First, we define the number of workers that evaluate trials concurrently. We also need to specify how long we want to run random search, by defining an upper limit on the total wall-clock time.

```
n_workers = 2  # Needs to be <= the number of available GPUs
max_wallclock_time = 12 * 60  # 12 minutes
```

The code for running ASHA is a simple variation of what we did for asynchronous random search.

```
mode = "min"
metric = "validation_error"
resource_attr = "epoch"

scheduler = ASHA(
    config_space,
    metric=metric,
    mode=mode,
    points_to_evaluate=[initial_config],
    max_resource_attr="max_epochs",
    resource_attr=resource_attr,
    grace_period=min_number_of_epochs,
    reduction_factor=eta,
)
```

```
INFO:syne_tune.optimizer.schedulers.fifo:max_resource_level = 10, as inferred␣
↪from config_space
INFO:syne_tune.optimizer.schedulers.fifo:Master random_seed = 3239342404
```

Here, `metric` and `resource_attr` specify the key names used with the `report` callback, and `max_resource_attr` denotes which input to the objective function corresponds to $r_{\max}$. Moreover, `grace_period` provides $r_{\min}$, and `reduction_factor` is $\eta$. We can run Syne Tune as before (this will take about 12 minutes):

```
trial_backend = PythonBackend(
    tune_function=hpo_objective_lenet_synetune,
    config_space=config_space,
)

stop_criterion = StoppingCriterion(max_wallclock_time=max_wallclock_time)
tuner = Tuner(
    trial_backend=trial_backend,
    scheduler=scheduler,
    stop_criterion=stop_criterion,
    n_workers=n_workers,
    print_update_interval=int(max_wallclock_time * 0.6),
)
tuner.run()
```

```
INFO:syne_tune.tuner:results of trials will be saved on /home/ubuntu/syne-tune/
↪python-entrypoint-2023-08-18-23-02-26-487
INFO:root:Detected 8 GPUs
INFO:root:running subprocess with command: /home/ubuntu/miniconda3/envs/np3/
↪bin/python /home/ubuntu/miniconda3/envs/np3/lib/python3.9/site-packages/syne_
```

```
↪tune/backend/python_backend/python_entrypoint.py --learning_rate 0.1 --batch_
↪size 128 --max_epochs 10 --tune_function_root /home/ubuntu/syne-tune/python-
↪entrypoint-2023-08-18-23-02-26-487/tune_function --tune_function_hash␣
↪836d63ee48b563fe3f22bde2d210ad89 --st_checkpoint_dir /home/ubuntu/syne-tune/
↪python-entrypoint-2023-08-18-23-02-26-487/0/checkpoints
INFO:syne_tune.tuner:(trial 0) - scheduled config {'learning_rate': 0.1,
↪'batch_size': 128, 'max_epochs': 10}
INFO:root:running subprocess with command: /home/ubuntu/miniconda3/envs/np3/
↪bin/python /home/ubuntu/miniconda3/envs/np3/lib/python3.9/site-packages/syne_
↪tune/backend/python_backend/python_entrypoint.py --learning_rate 0.
↪13853005130955554 --batch_size 121 --max_epochs 10 --tune_function_root /
↪home/ubuntu/syne-tune/python-entrypoint-2023-08-18-23-02-26-487/tune_
↪function --tune_function_hash 836d63ee48b563fe3f22bde2d210ad89 --st_
↪checkpoint_dir /home/ubuntu/syne-tune/python-entrypoint-2023-08-18-23-02-26-
↪487/1/checkpoints
INFO:syne_tune.tuner:(trial 1) - scheduled config {'learning_rate': 0.
↪13853005130955554, 'batch_size': 121, 'max_epochs': 10}
INFO:root:running subprocess with command: /home/ubuntu/miniconda3/envs/np3/
↪bin/python /home/ubuntu/miniconda3/envs/np3/lib/python3.9/site-packages/syne_
↪tune/backend/python_backend/python_entrypoint.py --learning_rate 0.
↪04888009860727095 --batch_size 226 --max_epochs 10 --tune_function_root /
↪home/ubuntu/syne-tune/python-entrypoint-2023-08-18-23-02-26-487/tune_
↪function --tune_function_hash 836d63ee48b563fe3f22bde2d210ad89 --st_
↪checkpoint_dir /home/ubuntu/syne-tune/python-entrypoint-2023-08-18-23-02-26-
↪487/2/checkpoints
INFO:syne_tune.tuner:(trial 2) - scheduled config {'learning_rate': 0.
↪04888009860727095, 'batch_size': 226, 'max_epochs': 10}
INFO:root:running subprocess with command: /home/ubuntu/miniconda3/envs/np3/
↪bin/python /home/ubuntu/miniconda3/envs/np3/lib/python3.9/site-packages/syne_
↪tune/backend/python_backend/python_entrypoint.py --learning_rate 0.
↪9214410391140626 --batch_size 246 --max_epochs 10 --tune_function_root /home/
↪ubuntu/syne-tune/python-entrypoint-2023-08-18-23-02-26-487/tune_function --
↪tune_function_hash 836d63ee48b563fe3f22bde2d210ad89 --st_checkpoint_dir /
↪home/ubuntu/syne-tune/python-entrypoint-2023-08-18-23-02-26-487/3/checkpoints
INFO:syne_tune.tuner:(trial 3) - scheduled config {'learning_rate': 0.
↪9214410391140626, 'batch_size': 246, 'max_epochs': 10}
INFO:syne_tune.tuner:Trial trial_id 1 completed.
INFO:root:running subprocess with command: /home/ubuntu/miniconda3/envs/np3/
↪bin/python /home/ubuntu/miniconda3/envs/np3/lib/python3.9/site-packages/syne_
↪tune/backend/python_backend/python_entrypoint.py --learning_rate 0.
↪018248573629167826 --batch_size 139 --max_epochs 10 --tune_function_root /
↪home/ubuntu/syne-tune/python-entrypoint-2023-08-18-23-02-26-487/tune_
↪function --tune_function_hash 836d63ee48b563fe3f22bde2d210ad89 --st_
↪checkpoint_dir /home/ubuntu/syne-tune/python-entrypoint-2023-08-18-23-02-26-
↪487/4/checkpoints
INFO:syne_tune.tuner:(trial 4) - scheduled config {'learning_rate': 0.
↪018248573629167826, 'batch_size': 139, 'max_epochs': 10}
INFO:root:running subprocess with command: /home/ubuntu/miniconda3/envs/np3/
↪bin/python /home/ubuntu/miniconda3/envs/np3/lib/python3.9/site-packages/syne_
↪tune/backend/python_backend/python_entrypoint.py --learning_rate 0.
↪02520991452721714 --batch_size 77 --max_epochs 10 --tune_function_root /home/
↪ubuntu/syne-tune/python-entrypoint-2023-08-18-23-02-26-487/tune_function --
↪tune_function_hash 836d63ee48b563fe3f22bde2d210ad89 --st_checkpoint_dir /
↪home/ubuntu/syne-tune/python-entrypoint-2023-08-18-23-02-26-487/5/checkpoints
INFO:syne_tune.tuner:(trial 5) - scheduled config {'learning_rate': 0.
```

```
↪02520991452721714, 'batch_size': 77, 'max_epochs': 10}
INFO:root:running subprocess with command: /home/ubuntu/miniconda3/envs/np3/
↪bin/python /home/ubuntu/miniconda3/envs/np3/lib/python3.9/site-packages/syne_
↪tune/backend/python_backend/python_entrypoint.py --learning_rate 0.
↪023767502921144167 --batch_size 235 --max_epochs 10 --tune_function_root /
↪home/ubuntu/syne-tune/python-entrypoint-2023-08-18-23-02-26-487/tune_
↪function --tune_function_hash 836d63ee48b563fe3f22bde2d210ad89 --st_
↪checkpoint_dir /home/ubuntu/syne-tune/python-entrypoint-2023-08-18-23-02-26-
↪487/6/checkpoints
INFO:syne_tune.tuner:(trial 6) - scheduled config {'learning_rate': 0.
↪023767502921144167, 'batch_size': 235, 'max_epochs': 10}
INFO:syne_tune.tuner:Trial trial_id 3 completed.
INFO:root:running subprocess with command: /home/ubuntu/miniconda3/envs/np3/
↪bin/python /home/ubuntu/miniconda3/envs/np3/lib/python3.9/site-packages/syne_
↪tune/backend/python_backend/python_entrypoint.py --learning_rate 0.
↪5128649019485186 --batch_size 133 --max_epochs 10 --tune_function_root /home/
↪ubuntu/syne-tune/python-entrypoint-2023-08-18-23-02-26-487/tune_function --
↪tune_function_hash 836d63ee48b563fe3f22bde2d210ad89 --st_checkpoint_dir /
↪home/ubuntu/syne-tune/python-entrypoint-2023-08-18-23-02-26-487/7/checkpoints
INFO:syne_tune.tuner:(trial 7) - scheduled config {'learning_rate': 0.
↪5128649019485186, 'batch_size': 133, 'max_epochs': 10}
INFO:root:running subprocess with command: /home/ubuntu/miniconda3/envs/np3/
↪bin/python /home/ubuntu/miniconda3/envs/np3/lib/python3.9/site-packages/syne_
↪tune/backend/python_backend/python_entrypoint.py --learning_rate 0.
↪0820355191393112 --batch_size 66 --max_epochs 10 --tune_function_root /home/
↪ubuntu/syne-tune/python-entrypoint-2023-08-18-23-02-26-487/tune_function --
↪tune_function_hash 836d63ee48b563fe3f22bde2d210ad89 --st_checkpoint_dir /
↪home/ubuntu/syne-tune/python-entrypoint-2023-08-18-23-02-26-487/8/checkpoints
INFO:syne_tune.tuner:(trial 8) - scheduled config {'learning_rate': 0.
↪0820355191393112, 'batch_size': 66, 'max_epochs': 10}
INFO:root:running subprocess with command: /home/ubuntu/miniconda3/envs/np3/
↪bin/python /home/ubuntu/miniconda3/envs/np3/lib/python3.9/site-packages/syne_
↪tune/backend/python_backend/python_entrypoint.py --learning_rate 0.
↪015979339095164306 --batch_size 236 --max_epochs 10 --tune_function_root /
↪home/ubuntu/syne-tune/python-entrypoint-2023-08-18-23-02-26-487/tune_
↪function --tune_function_hash 836d63ee48b563fe3f22bde2d210ad89 --st_
↪checkpoint_dir /home/ubuntu/syne-tune/python-entrypoint-2023-08-18-23-02-26-
↪487/9/checkpoints
INFO:syne_tune.tuner:(trial 9) - scheduled config {'learning_rate': 0.
↪015979339095164306, 'batch_size': 236, 'max_epochs': 10}
INFO:syne_tune.tuner:Trial trial_id 7 completed.
INFO:root:running subprocess with command: /home/ubuntu/miniconda3/envs/np3/
↪bin/python /home/ubuntu/miniconda3/envs/np3/lib/python3.9/site-packages/syne_
↪tune/backend/python_backend/python_entrypoint.py --learning_rate 0.
↪011287864128690532 --batch_size 128 --max_epochs 10 --tune_function_root /
↪home/ubuntu/syne-tune/python-entrypoint-2023-08-18-23-02-26-487/tune_
↪function --tune_function_hash 836d63ee48b563fe3f22bde2d210ad89 --st_
↪checkpoint_dir /home/ubuntu/syne-tune/python-entrypoint-2023-08-18-23-02-26-
↪487/10/checkpoints
INFO:syne_tune.tuner:(trial 10) - scheduled config {'learning_rate': 0.
↪011287864128690532, 'batch_size': 128, 'max_epochs': 10}
INFO:root:running subprocess with command: /home/ubuntu/miniconda3/envs/np3/
↪bin/python /home/ubuntu/miniconda3/envs/np3/lib/python3.9/site-packages/syne_
↪tune/backend/python_backend/python_entrypoint.py --learning_rate 0.
↪022375895483915522 --batch_size 80 --max_epochs 10 --tune_function_root /
```

```
↪home/ubuntu/syne-tune/python-entrypoint-2023-08-18-23-02-26-487/tune_
↪function --tune_function_hash 836d63ee48b563fe3f22bde2d210ad89 --st_
↪checkpoint_dir /home/ubuntu/syne-tune/python-entrypoint-2023-08-18-23-02-26-
↪487/11/checkpoints
INFO:syne_tune.tuner:(trial 11) - scheduled config {'learning_rate': 0.
↪022375895483915522, 'batch_size': 80, 'max_epochs': 10}
INFO:root:running subprocess with command: /home/ubuntu/miniconda3/envs/np3/
↪bin/python /home/ubuntu/miniconda3/envs/np3/lib/python3.9/site-packages/syne_
↪tune/backend/python_backend/python_entrypoint.py --learning_rate 0.
↪10869856278853329 --batch_size 246 --max_epochs 10 --tune_function_root /
↪home/ubuntu/syne-tune/python-entrypoint-2023-08-18-23-02-26-487/tune_
↪function --tune_function_hash 836d63ee48b563fe3f22bde2d210ad89 --st_
↪checkpoint_dir /home/ubuntu/syne-tune/python-entrypoint-2023-08-18-23-02-26-
↪487/12/checkpoints
INFO:syne_tune.tuner:(trial 12) - scheduled config {'learning_rate': 0.
↪10869856278853329, 'batch_size': 246, 'max_epochs': 10}
INFO:root:running subprocess with command: /home/ubuntu/miniconda3/envs/np3/
↪bin/python /home/ubuntu/miniconda3/envs/np3/lib/python3.9/site-packages/syne_
↪tune/backend/python_backend/python_entrypoint.py --learning_rate 0.
↪6321793222955815 --batch_size 142 --max_epochs 10 --tune_function_root /home/
↪ubuntu/syne-tune/python-entrypoint-2023-08-18-23-02-26-487/tune_function --
↪tune_function_hash 836d63ee48b563fe3f22bde2d210ad89 --st_checkpoint_dir /
↪home/ubuntu/syne-tune/python-entrypoint-2023-08-18-23-02-26-487/13/
↪checkpoints
INFO:syne_tune.tuner:(trial 13) - scheduled config {'learning_rate': 0.
↪6321793222955815, 'batch_size': 142, 'max_epochs': 10}
INFO:root:running subprocess with command: /home/ubuntu/miniconda3/envs/np3/
↪bin/python /home/ubuntu/miniconda3/envs/np3/lib/python3.9/site-packages/syne_
↪tune/backend/python_backend/python_entrypoint.py --learning_rate 0.
↪2536414155118574 --batch_size 249 --max_epochs 10 --tune_function_root /home/
↪ubuntu/syne-tune/python-entrypoint-2023-08-18-23-02-26-487/tune_function --
↪tune_function_hash 836d63ee48b563fe3f22bde2d210ad89 --st_checkpoint_dir /
↪home/ubuntu/syne-tune/python-entrypoint-2023-08-18-23-02-26-487/14/
↪checkpoints
INFO:syne_tune.tuner:(trial 14) - scheduled config {'learning_rate': 0.
↪2536414155118574, 'batch_size': 249, 'max_epochs': 10}
INFO:root:running subprocess with command: /home/ubuntu/miniconda3/envs/np3/
↪bin/python /home/ubuntu/miniconda3/envs/np3/lib/python3.9/site-packages/syne_
↪tune/backend/python_backend/python_entrypoint.py --learning_rate 0.
↪017930844776160398 --batch_size 152 --max_epochs 10 --tune_function_root /
↪home/ubuntu/syne-tune/python-entrypoint-2023-08-18-23-02-26-487/tune_
↪function --tune_function_hash 836d63ee48b563fe3f22bde2d210ad89 --st_
↪checkpoint_dir /home/ubuntu/syne-tune/python-entrypoint-2023-08-18-23-02-26-
↪487/15/checkpoints
INFO:syne_tune.tuner:(trial 15) - scheduled config {'learning_rate': 0.
↪017930844776160398, 'batch_size': 152, 'max_epochs': 10}
INFO:root:running subprocess with command: /home/ubuntu/miniconda3/envs/np3/
↪bin/python /home/ubuntu/miniconda3/envs/np3/lib/python3.9/site-packages/syne_
↪tune/backend/python_backend/python_entrypoint.py --learning_rate 0.
↪011532901021050378 --batch_size 207 --max_epochs 10 --tune_function_root /
↪home/ubuntu/syne-tune/python-entrypoint-2023-08-18-23-02-26-487/tune_
↪function --tune_function_hash 836d63ee48b563fe3f22bde2d210ad89 --st_
↪checkpoint_dir /home/ubuntu/syne-tune/python-entrypoint-2023-08-18-23-02-26-
↪487/16/checkpoints
INFO:syne_tune.tuner:(trial 16) - scheduled config {'learning_rate': 0.
```

```
↪011532901021050378, 'batch_size': 207, 'max_epochs': 10}
INFO:root:running subprocess with command: /home/ubuntu/miniconda3/envs/np3/
↪bin/python /home/ubuntu/miniconda3/envs/np3/lib/python3.9/site-packages/syne_
↪tune/backend/python_backend/python_entrypoint.py --learning_rate 0.
↪042449174869797324 --batch_size 210 --max_epochs 10 --tune_function_root /
↪home/ubuntu/syne-tune/python-entrypoint-2023-08-18-23-02-26-487/tune_
↪function --tune_function_hash 836d63ee48b563fe3f22bde2d210ad89 --st_
↪checkpoint_dir /home/ubuntu/syne-tune/python-entrypoint-2023-08-18-23-02-26-
↪487/17/checkpoints
INFO:syne_tune.tuner:(trial 17) - scheduled config {'learning_rate': 0.
↪042449174869797324, 'batch_size': 210, 'max_epochs': 10}
INFO:syne_tune.tuner:tuning status (last metric is reported)
 trial_id     status  iter  learning_rate  batch_size  max_epochs  epoch ␣
↪validation_error  worker-time
        0    Stopped     4       0.100000         128          10    4.0      ␣
↪   0.464695    37.212292
        1  Completed    10       0.138530         121          10   10.0      ␣
↪   0.253368    91.954761
        2    Stopped     2       0.048880         226          10    2.0      ␣
↪   0.901075    18.409374
        3  Completed    10       0.921441         246          10   10.0      ␣
↪   0.191130    91.572764
        4    Stopped     2       0.018249         139          10    2.0      ␣
↪   0.899994    19.555009
        5    Stopped     2       0.025210          77          10    2.0      ␣
↪   0.899966    26.743767
        6    Stopped     4       0.023768         235          10    4.0      ␣
↪   0.899760    38.330475
        7  Completed    10       0.512865         133          10   10.0      ␣
↪   0.231725    95.423449
        8    Stopped     4       0.082036          66          10    4.0      ␣
↪   0.376454    43.781020
        9    Stopped     2       0.015979         236          10    2.0      ␣
↪   0.899967    18.806864
       10    Stopped     4       0.011288         128          10    4.0      ␣
↪   0.899031    35.672628
       11    Stopped     2       0.022376          80          10    2.0      ␣
↪   0.900000    20.890393
       12    Stopped     2       0.108699         246          10    2.0      ␣
↪   0.900107    18.380533
       13    Stopped    10       0.632179         142          10   10.0      ␣
↪   0.180695    90.497290
       14    Stopped     4       0.253641         249          10    4.0      ␣
↪   0.425554    34.445888
       15    Stopped     2       0.017931         152          10    2.0      ␣
↪   0.899920    20.064630
       16 InProgress     1       0.011533         207          10    1.0      ␣
↪   0.899648    11.990970
       17 InProgress     0       0.042449         210          10      -      ␣
↪          -            -
2 trials running, 16 finished (3 until the end), 437.06s wallclock-time

INFO:root:running subprocess with command: /home/ubuntu/miniconda3/envs/np3/
↪bin/python /home/ubuntu/miniconda3/envs/np3/lib/python3.9/site-packages/syne_
↪tune/backend/python_backend/python_entrypoint.py --learning_rate 0.
```

```
↪5484553814219248 --batch_size 170 --max_epochs 10 --tune_function_root /home/
↪ubuntu/syne-tune/python-entrypoint-2023-08-18-23-02-26-487/tune_function --
↪tune_function_hash 836d63ee48b563fe3f22bde2d210ad89 --st_checkpoint_dir /
↪home/ubuntu/syne-tune/python-entrypoint-2023-08-18-23-02-26-487/18/
↪checkpoints
INFO:syne_tune.tuner:(trial 18) - scheduled config {'learning_rate': 0.
↪5484553814219248, 'batch_size': 170, 'max_epochs': 10}
INFO:root:running subprocess with command: /home/ubuntu/miniconda3/envs/np3/
↪bin/python /home/ubuntu/miniconda3/envs/np3/lib/python3.9/site-packages/syne_
↪tune/backend/python_backend/python_entrypoint.py --learning_rate 0.
↪011944500922210607 --batch_size 235 --max_epochs 10 --tune_function_root /
↪home/ubuntu/syne-tune/python-entrypoint-2023-08-18-23-02-26-487/tune_
↪function --tune_function_hash 836d63ee48b563fe3f22bde2d210ad89 --st_
↪checkpoint_dir /home/ubuntu/syne-tune/python-entrypoint-2023-08-18-23-02-26-
↪487/19/checkpoints
INFO:syne_tune.tuner:(trial 19) - scheduled config {'learning_rate': 0.
↪011944500922210607, 'batch_size': 235, 'max_epochs': 10}
INFO:root:running subprocess with command: /home/ubuntu/miniconda3/envs/np3/
↪bin/python /home/ubuntu/miniconda3/envs/np3/lib/python3.9/site-packages/syne_
↪tune/backend/python_backend/python_entrypoint.py --learning_rate 0.
↪08643278498726796 --batch_size 251 --max_epochs 10 --tune_function_root /
↪home/ubuntu/syne-tune/python-entrypoint-2023-08-18-23-02-26-487/tune_
↪function --tune_function_hash 836d63ee48b563fe3f22bde2d210ad89 --st_
↪checkpoint_dir /home/ubuntu/syne-tune/python-entrypoint-2023-08-18-23-02-26-
↪487/20/checkpoints
INFO:syne_tune.tuner:(trial 20) - scheduled config {'learning_rate': 0.
↪08643278498726796, 'batch_size': 251, 'max_epochs': 10}
INFO:root:running subprocess with command: /home/ubuntu/miniconda3/envs/np3/
↪bin/python /home/ubuntu/miniconda3/envs/np3/lib/python3.9/site-packages/syne_
↪tune/backend/python_backend/python_entrypoint.py --learning_rate 0.
↪7678647823352986 --batch_size 112 --max_epochs 10 --tune_function_root /home/
↪ubuntu/syne-tune/python-entrypoint-2023-08-18-23-02-26-487/tune_function --
↪tune_function_hash 836d63ee48b563fe3f22bde2d210ad89 --st_checkpoint_dir /
↪home/ubuntu/syne-tune/python-entrypoint-2023-08-18-23-02-26-487/21/
↪checkpoints
INFO:syne_tune.tuner:(trial 21) - scheduled config {'learning_rate': 0.
↪7678647823352986, 'batch_size': 112, 'max_epochs': 10}
INFO:syne_tune.tuner:Trial trial_id 18 completed.
INFO:root:running subprocess with command: /home/ubuntu/miniconda3/envs/np3/
↪bin/python /home/ubuntu/miniconda3/envs/np3/lib/python3.9/site-packages/syne_
↪tune/backend/python_backend/python_entrypoint.py --learning_rate 0.
↪25750421499886467 --batch_size 143 --max_epochs 10 --tune_function_root /
↪home/ubuntu/syne-tune/python-entrypoint-2023-08-18-23-02-26-487/tune_
↪function --tune_function_hash 836d63ee48b563fe3f22bde2d210ad89 --st_
↪checkpoint_dir /home/ubuntu/syne-tune/python-entrypoint-2023-08-18-23-02-26-
↪487/22/checkpoints
INFO:syne_tune.tuner:(trial 22) - scheduled config {'learning_rate': 0.
↪25750421499886467, 'batch_size': 143, 'max_epochs': 10}
INFO:root:running subprocess with command: /home/ubuntu/miniconda3/envs/np3/
↪bin/python /home/ubuntu/miniconda3/envs/np3/lib/python3.9/site-packages/syne_
↪tune/backend/python_backend/python_entrypoint.py --learning_rate 0.
↪04460413473978858 --batch_size 229 --max_epochs 10 --tune_function_root /
↪home/ubuntu/syne-tune/python-entrypoint-2023-08-18-23-02-26-487/tune_
↪function --tune_function_hash 836d63ee48b563fe3f22bde2d210ad89 --st_
↪checkpoint_dir /home/ubuntu/syne-tune/python-entrypoint-2023-08-18-23-02-26-
```

```
↪487/23/checkpoints
INFO:syne_tune.tuner:(trial 23) - scheduled config {'learning_rate': 0.
↪04460413473978858, 'batch_size': 229, 'max_epochs': 10}
INFO:root:running subprocess with command: /home/ubuntu/miniconda3/envs/np3/
↪bin/python /home/ubuntu/miniconda3/envs/np3/lib/python3.9/site-packages/syne_
↪tune/backend/python_backend/python_entrypoint.py --learning_rate 0.
↪03511449170254089 --batch_size 131 --max_epochs 10 --tune_function_root /
↪home/ubuntu/syne-tune/python-entrypoint-2023-08-18-23-02-26-487/tune_
↪function --tune_function_hash 836d63ee48b563fe3f22bde2d210ad89 --st_
↪checkpoint_dir /home/ubuntu/syne-tune/python-entrypoint-2023-08-18-23-02-26-
↪487/24/checkpoints
INFO:syne_tune.tuner:(trial 24) - scheduled config {'learning_rate': 0.
↪03511449170254089, 'batch_size': 131, 'max_epochs': 10}
INFO:root:running subprocess with command: /home/ubuntu/miniconda3/envs/np3/
↪bin/python /home/ubuntu/miniconda3/envs/np3/lib/python3.9/site-packages/syne_
↪tune/backend/python_backend/python_entrypoint.py --learning_rate 0.
↪013061175753480661 --batch_size 38 --max_epochs 10 --tune_function_root /
↪home/ubuntu/syne-tune/python-entrypoint-2023-08-18-23-02-26-487/tune_
↪function --tune_function_hash 836d63ee48b563fe3f22bde2d210ad89 --st_
↪checkpoint_dir /home/ubuntu/syne-tune/python-entrypoint-2023-08-18-23-02-26-
↪487/25/checkpoints
INFO:syne_tune.tuner:(trial 25) - scheduled config {'learning_rate': 0.
↪013061175753480661, 'batch_size': 38, 'max_epochs': 10}
INFO:root:running subprocess with command: /home/ubuntu/miniconda3/envs/np3/
↪bin/python /home/ubuntu/miniconda3/envs/np3/lib/python3.9/site-packages/syne_
↪tune/backend/python_backend/python_entrypoint.py --learning_rate 0.
↪028889846591178023 --batch_size 78 --max_epochs 10 --tune_function_root /
↪home/ubuntu/syne-tune/python-entrypoint-2023-08-18-23-02-26-487/tune_
↪function --tune_function_hash 836d63ee48b563fe3f22bde2d210ad89 --st_
↪checkpoint_dir /home/ubuntu/syne-tune/python-entrypoint-2023-08-18-23-02-26-
↪487/26/checkpoints
INFO:syne_tune.tuner:(trial 26) - scheduled config {'learning_rate': 0.
↪028889846591178023, 'batch_size': 78, 'max_epochs': 10}
INFO:root:running subprocess with command: /home/ubuntu/miniconda3/envs/np3/
↪bin/python /home/ubuntu/miniconda3/envs/np3/lib/python3.9/site-packages/syne_
↪tune/backend/python_backend/python_entrypoint.py --learning_rate 0.
↪06457802087000711 --batch_size 253 --max_epochs 10 --tune_function_root /
↪home/ubuntu/syne-tune/python-entrypoint-2023-08-18-23-02-26-487/tune_
↪function --tune_function_hash 836d63ee48b563fe3f22bde2d210ad89 --st_
↪checkpoint_dir /home/ubuntu/syne-tune/python-entrypoint-2023-08-18-23-02-26-
↪487/27/checkpoints
INFO:syne_tune.tuner:(trial 27) - scheduled config {'learning_rate': 0.
↪06457802087000711, 'batch_size': 253, 'max_epochs': 10}
INFO:root:running subprocess with command: /home/ubuntu/miniconda3/envs/np3/
↪bin/python /home/ubuntu/miniconda3/envs/np3/lib/python3.9/site-packages/syne_
↪tune/backend/python_backend/python_entrypoint.py --learning_rate 0.
↪7405331110191096 --batch_size 40 --max_epochs 10 --tune_function_root /home/
↪ubuntu/syne-tune/python-entrypoint-2023-08-18-23-02-26-487/tune_function --
↪tune_function_hash 836d63ee48b563fe3f22bde2d210ad89 --st_checkpoint_dir /
↪home/ubuntu/syne-tune/python-entrypoint-2023-08-18-23-02-26-487/28/
↪checkpoints
INFO:syne_tune.tuner:(trial 28) - scheduled config {'learning_rate': 0.
↪7405331110191096, 'batch_size': 40, 'max_epochs': 10}
INFO:syne_tune.stopping_criterion:reaching max wallclock time (720), stopping␣
↪there.
```

```
INFO:syne_tune.tuner:Stopping trials that may still be running.
INFO:syne_tune.tuner:Tuning finished, results of trials can be found on /home/
↪ubuntu/syne-tune/python-entrypoint-2023-08-18-23-02-26-487
--------------------
Resource summary (last result is reported):
 trial_id     status  iter  learning_rate  batch_size  max_epochs  epoch ␣
↪validation_error  worker-time
        0    Stopped     4       0.100000         128          10    4.0        ␣
↪     0.464695    37.212292
        1  Completed    10       0.138530         121          10   10.0        ␣
↪     0.253368    91.954761
        2    Stopped     2       0.048880         226          10    2.0        ␣
↪     0.901075    18.409374
        3  Completed    10       0.921441         246          10   10.0        ␣
↪     0.191130    91.572764
        4    Stopped     2       0.018249         139          10    2.0        ␣
↪     0.899994    19.555009
        5    Stopped     2       0.025210          77          10    2.0        ␣
↪     0.899966    26.743767
        6    Stopped     4       0.023768         235          10    4.0        ␣
↪     0.899760    38.330475
        7  Completed    10       0.512865         133          10   10.0        ␣
↪     0.231725    95.423449
        8    Stopped     4       0.082036          66          10    4.0        ␣
↪     0.376454    43.781020
        9    Stopped     2       0.015979         236          10    2.0        ␣
↪     0.899967    18.806864
       10    Stopped     4       0.011288         128          10    4.0        ␣
↪     0.899031    35.672628
       11    Stopped     2       0.022376          80          10    2.0        ␣
↪     0.900000    20.890393
       12    Stopped     2       0.108699         246          10    2.0        ␣
↪     0.900107    18.380533
       13    Stopped    10       0.632179         142          10   10.0        ␣
↪     0.180695    90.497290
       14    Stopped     4       0.253641         249          10    4.0        ␣
↪     0.425554    34.445888
       15    Stopped     2       0.017931         152          10    2.0        ␣
↪     0.899920    20.064630
       16    Stopped     2       0.011533         207          10    2.0        ␣
↪     0.900529    21.072846
       17    Stopped     4       0.042449         210          10    4.0        ␣
↪     0.899878    34.671282
       18  Completed    10       0.548455         170          10   10.0        ␣
↪     0.178600    92.981846
       19    Stopped     4       0.011945         235          10    4.0        ␣
↪     0.870940    35.437183
       20    Stopped     2       0.086433         251          10    2.0        ␣
↪     0.900077    18.594752
       21    Stopped    10       0.767865         112          10   10.0        ␣
↪     0.152976    95.257922
       22    Stopped     8       0.257504         143          10    8.0        ␣
↪     0.235383    75.945771
       23    Stopped     2       0.044604         229          10    2.0        ␣
↪     0.899916    19.311652
```

```
    24     Stopped      2        0.035114         131          10     2.0       0.900274     21.667166
    25     Stopped      4        0.013061          38          10     4.0       0.900319     58.163195
    26     Stopped      4        0.028890          78          10     4.0       0.900231     41.910056
    27  InProgress      0        0.064578         253          10       -              -            -
    28  InProgress      0        0.740533          40          10       -              -            -
2 trials running, 27 finished (4 until the end), 723.76s wallclock-time

validation_error: best 0.15019840002059937 for trial-id 21
--------------------
```

Note that we are running a variant of ASHA where underperforming trials are stopped early. This is different to our implementation in Section 19.4.1, where each training job is started with a fixed `max_epochs`. In the latter case, a well-performing trial which reaches the full 10 epochs, first needs to train 1, then 2, then 4, then 8 epochs, each time starting from scratch. This type of pause-and-resume scheduling can be implemented efficiently by checkpointing the training state after each epoch, but we avoid this extra complexity here. After the experiment has finished, we can retrieve and plot results.

```
d2l.set_figsize()
e = load_experiment(tuner.name)
e.plot()
```

```
WARNING:matplotlib.legend:No artists with labels found to put in legend.  Note that artists whose label start with an underscore are ignored when legend() is called with no argument.
```

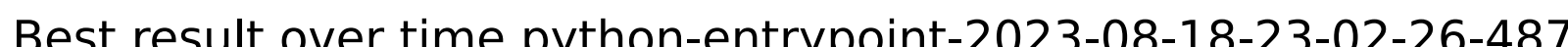

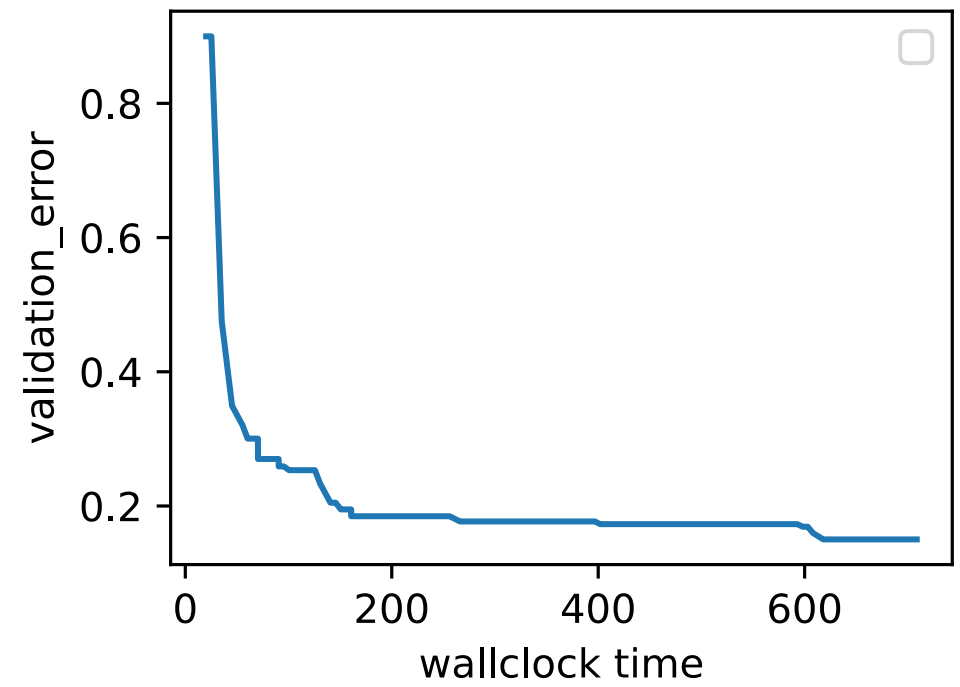

## 19.5.3 Visualize the Optimization Process

Once more, we visualize the learning curves of every trial (each color in the plot represents a trial). Compare this to asynchronous random search in Section 19.3. As we have seen for successive halving in Section 19.4, most of the trials are stopped at 1 or 2 epochs ($r_{\mathrm{min}}$ or $\eta * r_{\mathrm{min}}$). However, trials do not stop at the same point, because they require different amount of time per epoch. If we ran standard successive halving instead of ASHA, we would need to synchronize our workers, before we can promote configurations to the next rung level.

```
d2l.set_figsize([6, 2.5])
results = e.results
for trial_id in results.trial_id.unique():
    df = results[results["trial_id"] == trial_id]
    d2l.plt.plot(
        df["st_tuner_time"],
        df["validation_error"],
        marker="o"
    )
d2l.plt.xlabel("wall-clock time")
d2l.plt.ylabel("objective function")
```

```
Text(0, 0.5, 'objective function')
```

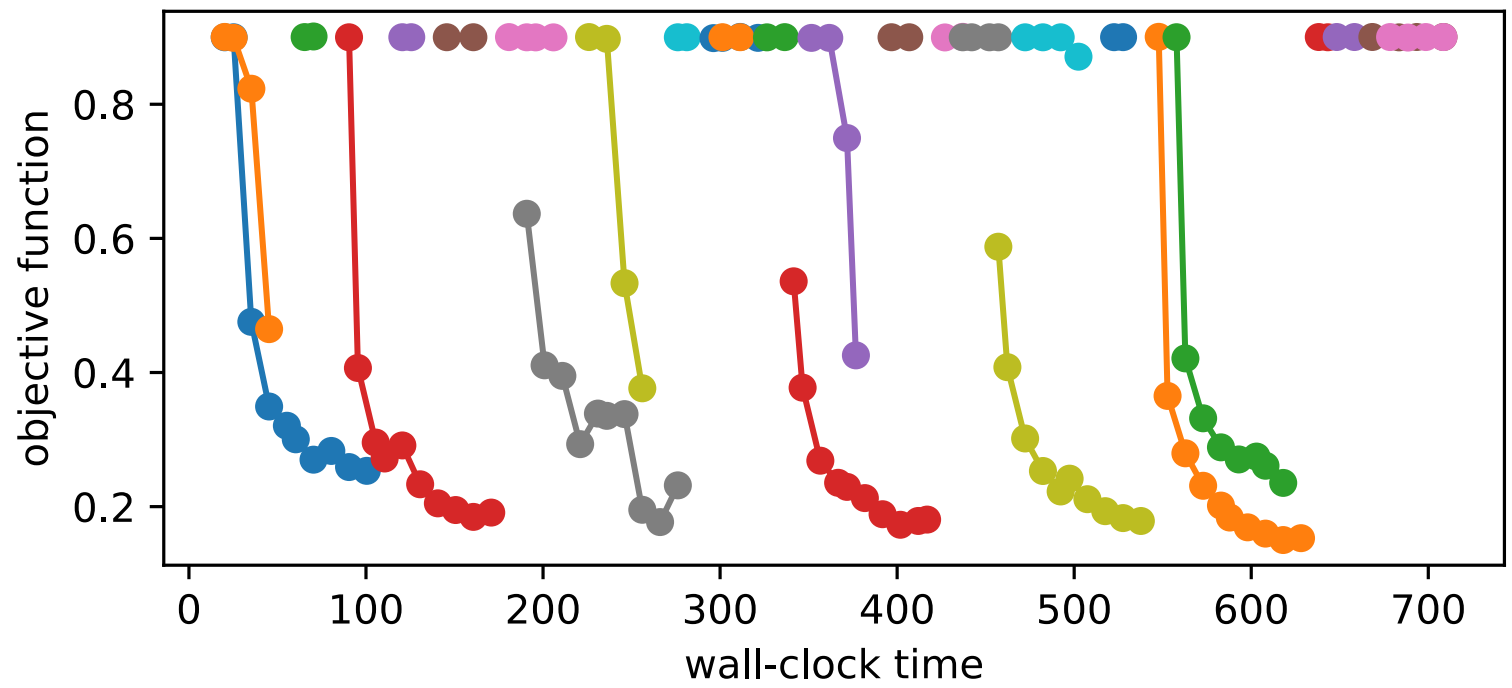

## 19.5.4 Summary

Compared to random search, successive halving is not quite as trivial to run in an asynchronous distributed setting. To avoid synchronisation points, we promote configurations as quickly as possible to the next rung level, even if this means promoting some wrong ones. In practice, this usually does not hurt much, and the gains of asynchronous versus synchronous scheduling are usually much higher than the loss of the suboptimal decision making.

Discussions[273].

# 20 Generative Adversarial Networks

## 20.1 Generative Adversarial Networks

Throughout most of this book, we have talked about how to make predictions. In some form or another, we used deep neural networks to learn mappings from data examples to labels. This kind of learning is called discriminative learning, as in, we'd like to be able to discriminate between photos of cats and photos of dogs. Classifiers and regressors are both examples of discriminative learning. And neural networks trained by backpropagation have upended everything we thought we knew about discriminative learning on large complicated datasets. Classification accuracies on high-res images have gone from useless to human-level (with some caveats) in just 5-6 years. We will spare you another spiel about all the other discriminative tasks where deep neural networks do astoundingly well.

But there is more to machine learning than just solving discriminative tasks. For example, given a large dataset, without any labels, we might want to learn a model that concisely captures the characteristics of this data. Given such a model, we could sample synthetic data examples that resemble the distribution of the training data. For example, given a large corpus of photographs of faces, we might want to be able to generate a new photorealistic image that looks like it might plausibly have come from the same dataset. This kind of learning is called generative modeling.

Until recently, we had no method that could synthesize novel photorealistic images. But the success of deep neural networks for discriminative learning opened up new possibilities. One big trend over the last three years has been the application of discriminative deep nets to overcome challenges in problems that we do not generally think of as supervised learning problems. The recurrent neural network language models are one example of using a discriminative network (trained to predict the next character) that once trained can act as a generative model.

In 2014, a breakthrough paper introduced Generative adversarial networks (GANs) (Goodfellow *et al.*, 2014), a clever new way to leverage the power of discriminative models to get good generative models. At their heart, GANs rely on the idea that a data generator is good if we cannot tell fake data apart from real data. In statistics, this is called a two-sample test - a test to answer the question whether datasets $X = \{x_1, \ldots, x_n\}$ and $X' = \{x'_1, \ldots, x'_n\}$ were drawn from the same distribution. The main difference between most statistics papers and GANs is that the latter use this idea in a constructive way. In other words, rather than just training a model to say "hey, these two datasets do not look like they came from the same distribution", they use the two-sample test[274] to provide training signals to a generative model. This allows us to improve the data generator until it generates something that resembles the real data. At the very least, it needs to fool the classifier even if our classifier is a state of the art deep neural network.

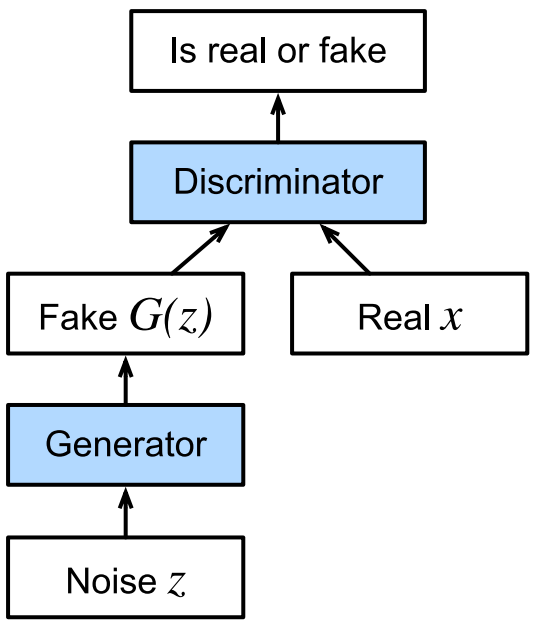

Fig. 20.1.1 Generative Adversarial Networks

The GAN architecture is illustrated in Fig. 20.1.1. As you can see, there are two pieces in GAN architecture - first off, we need a device (say, a deep network but it really could be anything, such as a game rendering engine) that might potentially be able to generate data that looks just like the real thing. If we are dealing with images, this needs to generate images. If we are dealing with speech, it needs to generate audio sequences, and so on. We call this the generator network. The second component is the discriminator network. It attempts to distinguish fake and real data from each other. Both networks are in competition with each other. The generator network attempts to fool the discriminator network. At that point, the discriminator network adapts to the new fake data. This information, in turn is used to improve the generator network, and so on.

The discriminator is a binary classifier to distinguish if the input $x$ is real (from real data) or fake (from the generator). Typically, the discriminator outputs a scalar prediction $o \in \mathbb{R}$ for input $\mathbf{x}$, such as using a fully connected layer with hidden size 1, and then applies sigmoid function to obtain the predicted probability $D(\mathbf{x}) = 1/(1+e^{-o})$. Assume the label $y$ for the true data is 1 and 0 for the fake data. We train the discriminator to minimize the cross-entropy loss, *i.e.*,

$$\min_D \{ - y \log D(\mathbf{x}) - (1-y)\log(1-D(\mathbf{x})) \}, \tag{20.1.1}$$

For the generator, it first draws some parameter $\mathbf{z} \in \mathbb{R}^d$ from a source of randomness, *e.g.*, a normal distribution $\mathbf{z} \sim \mathcal{N}(0,1)$. We often call $\mathbf{z}$ as the latent variable. It then applies a function to generate $\mathbf{x}' = G(\mathbf{z})$. The goal of the generator is to fool the discriminator to classify $\mathbf{x}' = G(\mathbf{z})$ as true data, *i.e.*, we want $D(G(\mathbf{z})) \approx 1$. In other words, for a given discriminator $D$, we update the parameters of the generator $G$ to maximize the cross-entropy loss when $y = 0$, *i.e.*,

$$\max_G \{ - (1-y) \log(1-D(G(\mathbf{z}))) \} = \max_G \{ - \log(1-D(G(\mathbf{z}))) \}. \tag{20.1.2}$$

If the generator does a perfect job, then $D(\mathbf{x}') \approx 1$, so the above loss is near 0, which results in the gradients that are too small to make good progress for the discriminator. So commonly, we minimize the following loss:

$$\min_G \{-y \log(D(G(\mathbf{z})))\} = \min_G \{-\log(D(G(\mathbf{z})))\}, \tag{20.1.3}$$

which is just feeding $\mathbf{x}' = G(\mathbf{z})$ into the discriminator but giving label $y = 1$.

To sum up, $D$ and $G$ are playing a "minimax" game with the comprehensive objective function:

$$\min_D \max_G \{-E_{x\sim \text{Data}} \log D(\mathbf{x}) - E_{z\sim \text{Noise}} \log(1 - D(G(\mathbf{z})))\}. \tag{20.1.4}$$

Many of the GANs applications are in the context of images. As a demonstration purpose, we are going to content ourselves with fitting a much simpler distribution first. We will illustrate what happens if we use GANs to build the world's most inefficient estimator of parameters for a Gaussian. Let's get started.

```
%matplotlib inline
import torch
from torch import nn
from d2l import torch as d2l
```

## 20.1.1 Generate Some "Real" Data

Since this is going to be the world's lamest example, we simply generate data drawn from a Gaussian.

```
X = torch.normal(0.0, 1, (1000, 2))
A = torch.tensor([[1, 2], [-0.1, 0.5]])
b = torch.tensor([1, 2])
data = torch.matmul(X, A) + b
```

Let's see what we got. This should be a Gaussian shifted in some rather arbitrary way with mean $b$ and covariance matrix $A^T A$.

```
d2l.set_figsize()
d2l.plt.scatter(data[:100, 0].detach().numpy(), data[:100, 1].detach().
↪numpy());
print(f'The covariance matrix is\n{torch.matmul(A.T, A)}')
```

```
The covariance matrix is
tensor([[1.0100, 1.9500],
        [1.9500, 4.2500]])
```

```
batch_size = 8
data_iter = d2l.load_array((data,), batch_size)
```

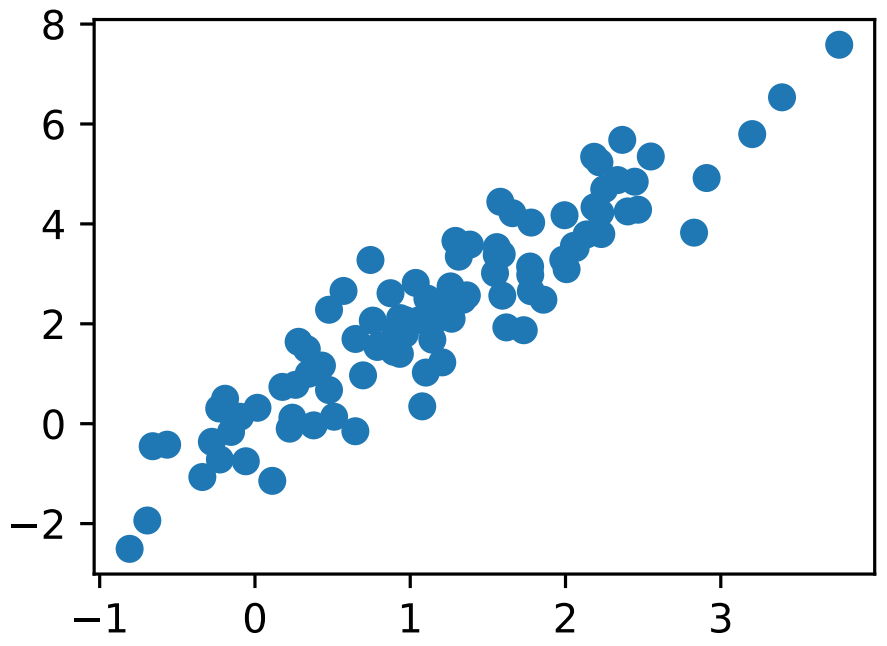

## 20.1.2 Generator

Our generator network will be the simplest network possible - a single layer linear model. This is since we will be driving that linear network with a Gaussian data generator. Hence, it literally only needs to learn the parameters to fake things perfectly.

```
net_G = nn.Sequential(nn.Linear(2, 2))
```

## 20.1.3 Discriminator

For the discriminator we will be a bit more discriminating: we will use an MLP with 3 layers to make things a bit more interesting.

```
net_D = nn.Sequential(
    nn.Linear(2, 5), nn.Tanh(),
    nn.Linear(5, 3), nn.Tanh(),
    nn.Linear(3, 1))
```

## 20.1.4 Training

First we define a function to update the discriminator.

```
#@save
def update_D(X, Z, net_D, net_G, loss, trainer_D):
    """Update discriminator."""
    batch_size = X.shape[0]
    ones = torch.ones((batch_size,), device=X.device)
    zeros = torch.zeros((batch_size,), device=X.device)
    trainer_D.zero_grad()
    real_Y = net_D(X)
    fake_X = net_G(Z)
    # Do not need to compute gradient for `net_G`, detach it from
    # computing gradients.
```

```
    fake_Y = net_D(fake_X.detach())
    loss_D = (loss(real_Y, ones.reshape(real_Y.shape)) +
              loss(fake_Y, zeros.reshape(fake_Y.shape))) / 2
    loss_D.backward()
    trainer_D.step()
    return loss_D
```

The generator is updated similarly. Here we reuse the cross-entropy loss but change the label of the fake data from 0 to 1.

```
#@save
def update_G(Z, net_D, net_G, loss, trainer_G):
    """Update generator."""
    batch_size = Z.shape[0]
    ones = torch.ones((batch_size,), device=Z.device)
    trainer_G.zero_grad()
    # We could reuse `fake_X` from `update_D` to save computation
    fake_X = net_G(Z)
    # Recomputing `fake_Y` is needed since `net_D` is changed
    fake_Y = net_D(fake_X)
    loss_G = loss(fake_Y, ones.reshape(fake_Y.shape))
    loss_G.backward()
    trainer_G.step()
    return loss_G
```

Both the discriminator and the generator performs a binary logistic regression with the cross-entropy loss. We use Adam to smooth the training process. In each iteration, we first update the discriminator and then the generator. We visualize both losses and generated examples.

```
def train(net_D, net_G, data_iter, num_epochs, lr_D, lr_G, latent_dim, data):
    loss = nn.BCEWithLogitsLoss(reduction='sum')
    for w in net_D.parameters():
        nn.init.normal_(w, 0, 0.02)
    for w in net_G.parameters():
        nn.init.normal_(w, 0, 0.02)
    trainer_D = torch.optim.Adam(net_D.parameters(), lr=lr_D)
    trainer_G = torch.optim.Adam(net_G.parameters(), lr=lr_G)
    animator = d2l.Animator(xlabel='epoch', ylabel='loss',
                            xlim=[1, num_epochs], nrows=2, figsize=(5, 5),
                            legend=['discriminator', 'generator'])
    animator.fig.subplots_adjust(hspace=0.3)
    for epoch in range(num_epochs):
        # Train one epoch
        timer = d2l.Timer()
        metric = d2l.Accumulator(3)  # loss_D, loss_G, num_examples
        for (X,) in data_iter:
            batch_size = X.shape[0]
            Z = torch.normal(0, 1, size=(batch_size, latent_dim))
            metric.add(update_D(X, Z, net_D, net_G, loss, trainer_D),
                       update_G(Z, net_D, net_G, loss, trainer_G),
                       batch_size)
```

(continued from previous page)

```
        # Visualize generated examples
        Z = torch.normal(0, 1, size=(100, latent_dim))
        fake_X = net_G(Z).detach().numpy()
        animator.axes[1].cla()
        animator.axes[1].scatter(data[:, 0], data[:, 1])
        animator.axes[1].scatter(fake_X[:, 0], fake_X[:, 1])
        animator.axes[1].legend(['real', 'generated'])
        # Show the losses
        loss_D, loss_G = metric[0]/metric[2], metric[1]/metric[2]
        animator.add(epoch + 1, (loss_D, loss_G))
    print(f'loss_D {loss_D:.3f}, loss_G {loss_G:.3f}, '
          f'{metric[2] / timer.stop():.1f} examples/sec')
```

Now we specify the hyperparameters to fit the Gaussian distribution.

```
lr_D, lr_G, latent_dim, num_epochs = 0.05, 0.005, 2, 20
train(net_D, net_G, data_iter, num_epochs, lr_D, lr_G,
      latent_dim, data[:100].detach().numpy())
```

```
loss_D 0.693, loss_G 0.693, 985.2 examples/sec
```

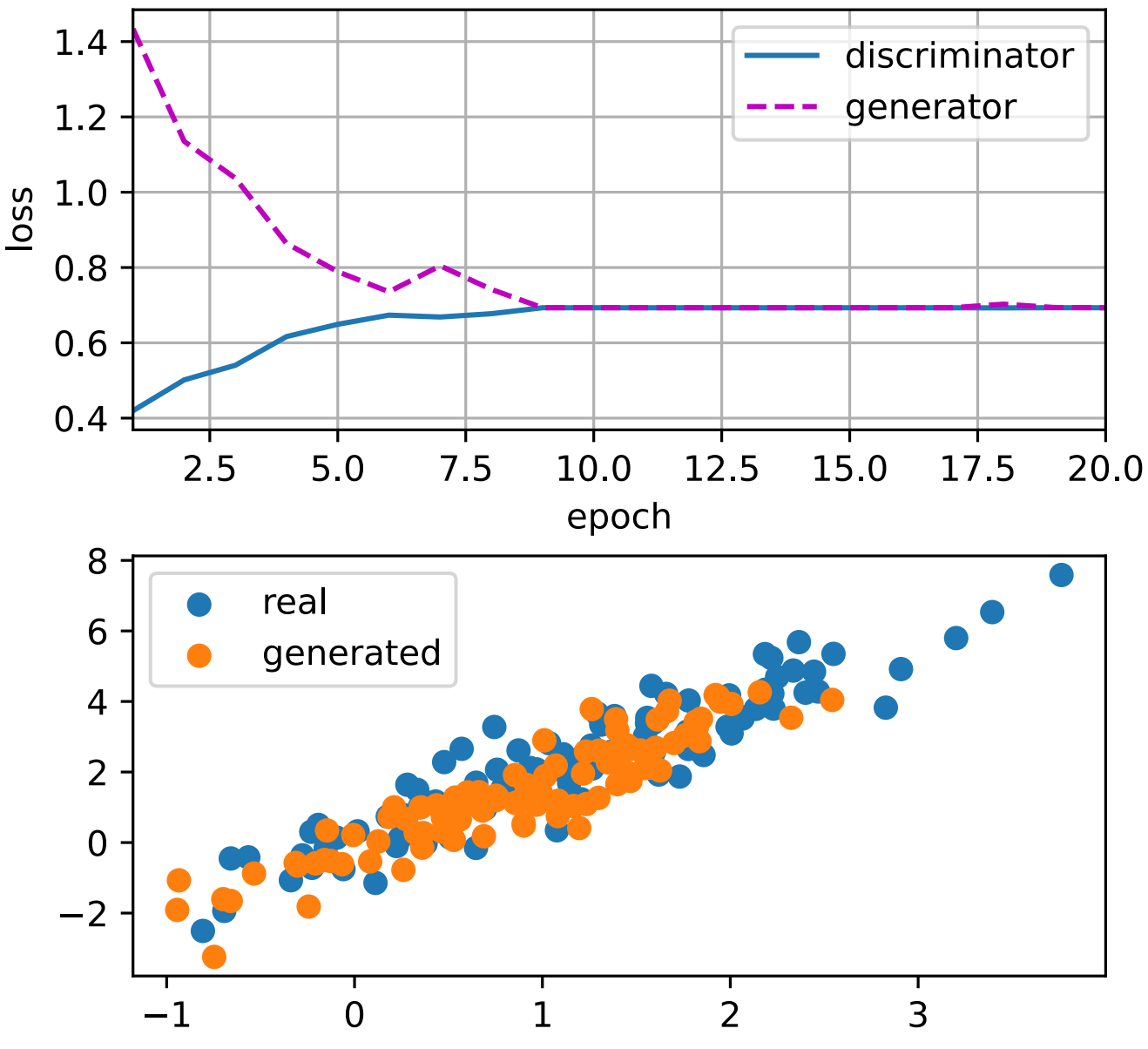

## 20.1.5 Summary

- Generative adversarial networks (GANs) composes of two deep networks, the generator and the discriminator.

- The generator generates the image as much closer to the true image as possible to fool the discriminator, via maximizing the cross-entropy loss, *i.e.*, $\max \log(D(\mathbf{x}'))$.
- The discriminator tries to distinguish the generated images from the true images, via minimizing the cross-entropy loss, *i.e.*, $\min -y \log D(\mathbf{x}) - (1-y) \log(1 - D(\mathbf{x}))$.

### 20.1.6 Exercises

- Does an equilibrium exist where the generator wins, *i.e.* the discriminator ends up unable to distinguish the two distributions on finite samples?

Discussions[275].

## 20.2 Deep Convolutional Generative Adversarial Networks

In Section 20.1, we introduced the basic ideas behind how GANs work. We showed that they can draw samples from some simple, easy-to-sample distribution, like a uniform or normal distribution, and transform them into samples that appear to match the distribution of some dataset. And while our example of matching a 2D Gaussian distribution got the point across, it is not especially exciting.

In this section, we will demonstrate how you can use GANs to generate photorealistic images. We will be basing our models on the deep convolutional GANs (DCGAN) introduced in Radford *et al.* (2015). We will borrow the convolutional architecture that have proven so successful for discriminative computer vision problems and show how via GANs, they can be leveraged to generate photorealistic images.

```
import warnings
import torch
import torchvision
from torch import nn
from d2l import torch as d2l
```

### 20.2.1 The Pokemon Dataset

The dataset we will use is a collection of Pokemon sprites obtained from pokemondb[276]. First download, extract and load this dataset.

```
#@save
d2l.DATA_HUB['pokemon'] = (d2l.DATA_URL + 'pokemon.zip',
                           'c065c0e2593b8b161a2d7873e42418bf6a21106c')

data_dir = d2l.download_extract('pokemon')
pokemon = torchvision.datasets.ImageFolder(data_dir)
```

We resize each image into $64 \times 64$. The ToTensor transformation will project the pixel value into $[0, 1]$, while our generator will use the tanh function to obtain outputs in $[-1, 1]$. Therefore we normalize the data with $0.5$ mean and $0.5$ standard deviation to match the value range.

```
batch_size = 256
transformer = torchvision.transforms.Compose([
    torchvision.transforms.Resize((64, 64)),
    torchvision.transforms.ToTensor(),
    torchvision.transforms.Normalize(0.5, 0.5)
])
pokemon.transform = transformer
data_iter = torch.utils.data.DataLoader(
    pokemon, batch_size=batch_size,
    shuffle=True, num_workers=d2l.get_dataloader_workers())
```

Let's visualize the first 20 images.

```
warnings.filterwarnings('ignore')
d2l.set_figsize((4, 4))
for X, y in data_iter:
    imgs = X[:20,:,:,:].permute(0, 2, 3, 1)/2+0.5
    d2l.show_images(imgs, num_rows=4, num_cols=5)
    break
```

### 20.2.2 The Generator

The generator needs to map the noise variable $\mathbf{z} \in \mathbb{R}^d$, a length-$d$ vector, to a RGB image with width and height to be $64 \times 64$ . In Section 14.11 we introduced the fully convolutional network that uses transposed convolution layer (refer to Section 14.10) to enlarge input size. The basic block of the generator contains a transposed convolution layer followed by the batch normalization and ReLU activation.

```
class G_block(nn.Module):
    def __init__(self, out_channels, in_channels=3, kernel_size=4, strides=2,
                 padding=1, **kwargs):
        super(G_block, self).__init__(**kwargs)
        self.conv2d_trans = nn.ConvTranspose2d(in_channels, out_channels,
                                kernel_size, strides, padding, bias=False)
        self.batch_norm = nn.BatchNorm2d(out_channels)
        self.activation = nn.ReLU()
```

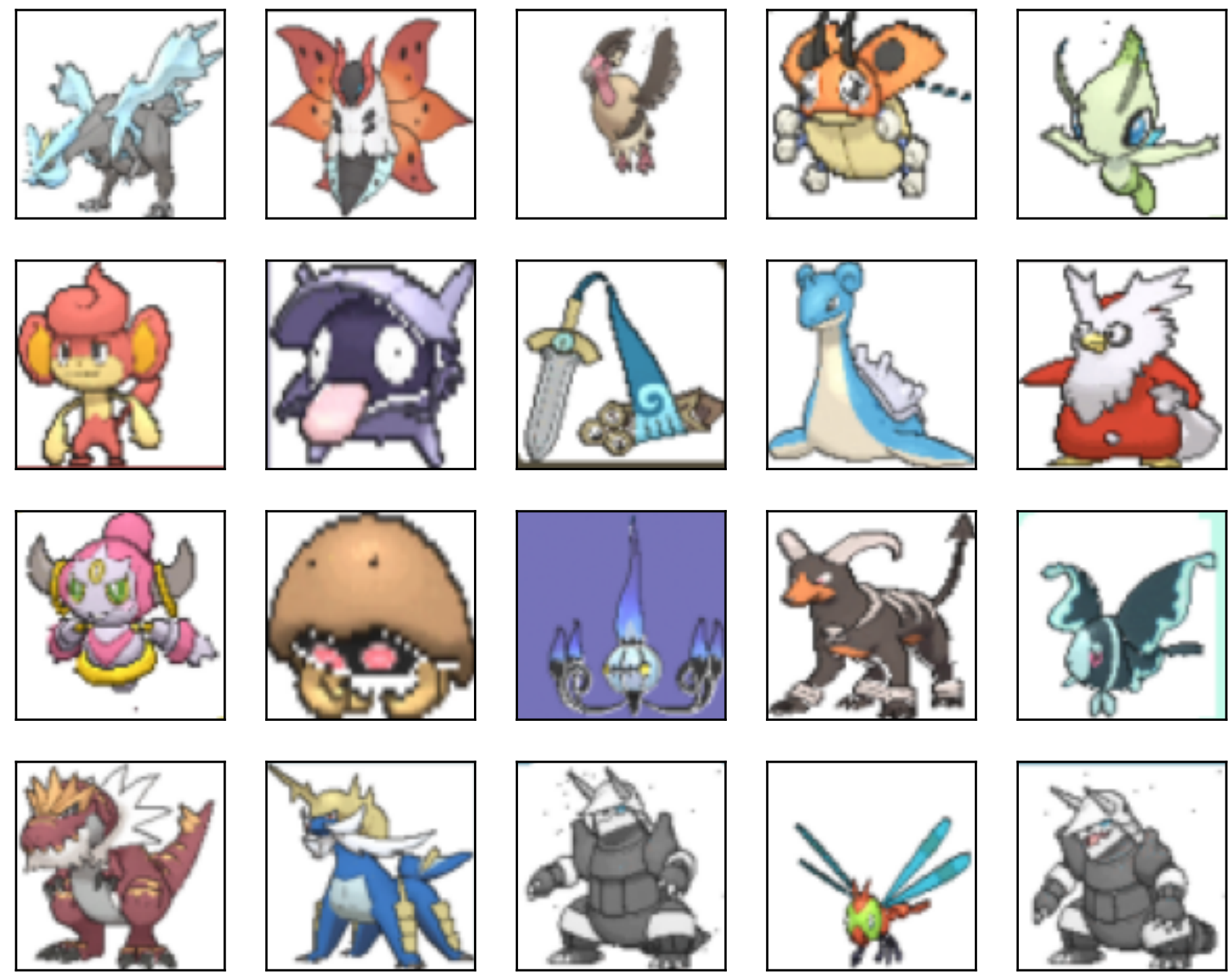

(continued from previous page)

```
    def forward(self, X):
        return self.activation(self.batch_norm(self.conv2d_trans(X)))
```

In default, the transposed convolution layer uses a $k_h = k_w = 4$ kernel, a $s_h = s_w = 2$ strides, and a $p_h = p_w = 1$ padding. With a input shape of $n_h^{'} \times n_w^{'} = 16 \times 16$, the generator block will double input's width and height.

$$\begin{aligned} n_h^{'} \times n_w^{'} &= [(n_h k_h - (n_h - 1)(k_h - s_h) - 2p_h] \times [(n_w k_w - (n_w - 1)(k_w - s_w) - 2p_w] \\ &= [(k_h + s_h(n_h - 1) - 2p_h] \times [(k_w + s_w(n_w - 1) - 2p_w] \\ &= [(4 + 2 \times (16 - 1) - 2 \times 1] \times [(4 + 2 \times (16 - 1) - 2 \times 1] \\ &= 32 \times 32. \end{aligned} \tag{20.2.1}$$

```
x = torch.zeros((2, 3, 16, 16))
g_blk = G_block(20)
g_blk(x).shape
```

```
torch.Size([2, 20, 32, 32])
```

If changing the transposed convolution layer to a $4 \times 4$ kernel, $1 \times 1$ strides and zero padding.

With a input size of $1 \times 1$, the output will have its width and height increased by 3 respectively.

```
x = torch.zeros((2, 3, 1, 1))
g_blk = G_block(20, strides=1, padding=0)
g_blk(x).shape
```

```
torch.Size([2, 20, 4, 4])
```

The generator consists of four basic blocks that increase input's both width and height from 1 to 32. At the same time, it first projects the latent variable into $64 \times 8$ channels, and then halve the channels each time. At last, a transposed convolution layer is used to generate the output. It further doubles the width and height to match the desired $64 \times 64$ shape, and reduces the channel size to 3. The tanh activation function is applied to project output values into the $(-1, 1)$ range.

```
n_G = 64
net_G = nn.Sequential(
    G_block(in_channels=100, out_channels=n_G*8,
            strides=1, padding=0),                  # Output: (64 * 8, 4, 4)
    G_block(in_channels=n_G*8, out_channels=n_G*4), # Output: (64 * 4, 8, 8)
    G_block(in_channels=n_G*4, out_channels=n_G*2), # Output: (64 * 2, 16, 16)
    G_block(in_channels=n_G*2, out_channels=n_G),   # Output: (64, 32, 32)
    nn.ConvTranspose2d(in_channels=n_G, out_channels=3,
                       kernel_size=4, stride=2, padding=1, bias=False),
    nn.Tanh())  # Output: (3, 64, 64)
```

Generate a 100 dimensional latent variable to verify the generator's output shape.

```
x = torch.zeros((1, 100, 1, 1))
net_G(x).shape
```

```
torch.Size([1, 3, 64, 64])
```

### 20.2.3 Discriminator

The discriminator is a normal convolutional network network except that it uses a leaky ReLU as its activation function. Given $\alpha \in [0, 1]$, its definition is

$$\text{leaky ReLU}(x) = \begin{cases} x & \text{if } x > 0 \\ \alpha x & \text{otherwise} \end{cases}. \tag{20.2.2}$$

As it can be seen, it is normal ReLU if $\alpha = 0$, and an identity function if $\alpha = 1$. For $\alpha \in (0, 1)$, leaky ReLU is a nonlinear function that give a non-zero output for a negative input. It aims to fix the "dying ReLU" problem that a neuron might always output a negative value and therefore cannot make any progress since the gradient of ReLU is 0.

```
alphas = [0, .2, .4, .6, .8, 1]
x = torch.arange(-2, 1, 0.1)
Y = [nn.LeakyReLU(alpha)(x).detach().numpy() for alpha in alphas]
d2l.plot(x.detach().numpy(), Y, 'x', 'y', alphas)
```

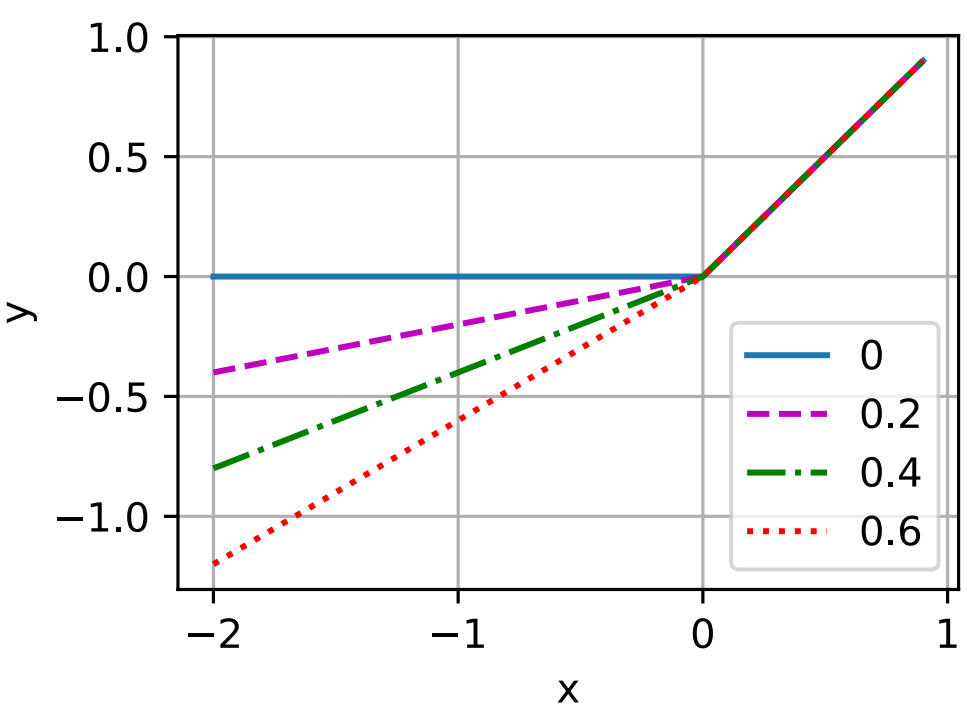

The basic block of the discriminator is a convolution layer followed by a batch normalization layer and a leaky ReLU activation. The hyperparameters of the convolution layer are similar to the transpose convolution layer in the generator block.

```
class D_block(nn.Module):
    def __init__(self, out_channels, in_channels=3, kernel_size=4, strides=2,
                 padding=1, alpha=0.2, **kwargs):
        super(D_block, self).__init__(**kwargs)
        self.conv2d = nn.Conv2d(in_channels, out_channels, kernel_size,
                                strides, padding, bias=False)
        self.batch_norm = nn.BatchNorm2d(out_channels)
        self.activation = nn.LeakyReLU(alpha, inplace=True)

    def forward(self, X):
        return self.activation(self.batch_norm(self.conv2d(X)))
```

A basic block with default settings will halve the width and height of the inputs, as we demonstrated in Section 7.3. For example, given a input shape $n_h = n_w = 16$, with a kernel shape $k_h = k_w = 4$, a stride shape $s_h = s_w = 2$, and a padding shape $p_h = p_w = 1$, the output shape will be:

$$\begin{aligned} n_h' \times n_w' &= \lfloor (n_h - k_h + 2p_h + s_h)/s_h \rfloor \times \lfloor (n_w - k_w + 2p_w + s_w)/s_w \rfloor \\ &= \lfloor (16 - 4 + 2 \times 1 + 2)/2 \rfloor \times \lfloor (16 - 4 + 2 \times 1 + 2)/2 \rfloor \\ &= 8 \times 8. \end{aligned} \tag{20.2.3}$$

```
x = torch.zeros((2, 3, 16, 16))
d_blk = D_block(20)
d_blk(x).shape
```

```
torch.Size([2, 20, 8, 8])
```

The discriminator is a mirror of the generator.

```
n_D = 64
net_D = nn.Sequential(
    D_block(n_D),  # Output: (64, 32, 32)
    D_block(in_channels=n_D, out_channels=n_D*2),  # Output: (64 * 2, 16, 16)
    D_block(in_channels=n_D*2, out_channels=n_D*4),  # Output: (64 * 4, 8, 8)
    D_block(in_channels=n_D*4, out_channels=n_D*8),  # Output: (64 * 8, 4, 4)
    nn.Conv2d(in_channels=n_D*8, out_channels=1,
              kernel_size=4, bias=False))  # Output: (1, 1, 1)
```

It uses a convolution layer with output channel 1 as the last layer to obtain a single prediction value.

```
x = torch.zeros((1, 3, 64, 64))
net_D(x).shape
```

```
torch.Size([1, 1, 1, 1])
```

### 20.2.4 Training

Compared to the basic GAN in Section 20.1, we use the same learning rate for both generator and discriminator since they are similar to each other. In addition, we change $\beta_1$ in Adam (Section 12.10) from $0.9$ to $0.5$. It decreases the smoothness of the momentum, the exponentially weighted moving average of past gradients, to take care of the rapid changing gradients because the generator and the discriminator fight with each other. Besides, the random generated noise Z, is a 4-D tensor and we are using GPU to accelerate the computation.

```
def train(net_D, net_G, data_iter, num_epochs, lr, latent_dim,
          device=d2l.try_gpu()):
    loss = nn.BCEWithLogitsLoss(reduction='sum')
    for w in net_D.parameters():
        nn.init.normal_(w, 0, 0.02)
    for w in net_G.parameters():
        nn.init.normal_(w, 0, 0.02)
    net_D, net_G = net_D.to(device), net_G.to(device)
    trainer_hp = {'lr': lr, 'betas': [0.5,0.999]}
    trainer_D = torch.optim.Adam(net_D.parameters(), **trainer_hp)
    trainer_G = torch.optim.Adam(net_G.parameters(), **trainer_hp)
    animator = d2l.Animator(xlabel='epoch', ylabel='loss',
                            xlim=[1, num_epochs], nrows=2, figsize=(5, 5),
                            legend=['discriminator', 'generator'])
    animator.fig.subplots_adjust(hspace=0.3)
    for epoch in range(1, num_epochs + 1):
        # Train one epoch
```

```
        timer = d2l.Timer()
        metric = d2l.Accumulator(3)  # loss_D, loss_G, num_examples
        for X, _ in data_iter:
            batch_size = X.shape[0]
            Z = torch.normal(0, 1, size=(batch_size, latent_dim, 1, 1))
            X, Z = X.to(device), Z.to(device)
            metric.add(d2l.update_D(X, Z, net_D, net_G, loss, trainer_D),
                       d2l.update_G(Z, net_D, net_G, loss, trainer_G),
                       batch_size)
        # Show generated examples
        Z = torch.normal(0, 1, size=(21, latent_dim, 1, 1), device=device)
        # Normalize the synthetic data to N(0, 1)
        fake_x = net_G(Z).permute(0, 2, 3, 1) / 2 + 0.5
        imgs = torch.cat(
            [torch.cat([
                fake_x[i * 7 + j].cpu().detach() for j in range(7)], dim=1)
             for i in range(len(fake_x)//7)], dim=0)
        animator.axes[1].cla()
        animator.axes[1].imshow(imgs)
        # Show the losses
        loss_D, loss_G = metric[0] / metric[2], metric[1] / metric[2]
        animator.add(epoch, (loss_D, loss_G))
    print(f'loss_D {loss_D:.3f}, loss_G {loss_G:.3f}, '
          f'{metric[2] / timer.stop():.1f} examples/sec on {str(device)}')
```

We train the model with a small number of epochs just for demonstration. For better performance, the variable num_epochs can be set to a larger number.

```
latent_dim, lr, num_epochs = 100, 0.005, 20
train(net_D, net_G, data_iter, num_epochs, lr, latent_dim)
```

```
loss_D 0.085, loss_G 7.452, 1912.6 examples/sec on cuda:0
```

### 20.2.5 Summary

- DCGAN architecture has four convolutional layers for the Discriminator and four "fractionally-strided" convolutional layers for the Generator.
- The Discriminator is a 4-layer strided convolutions with batch normalization (except its input layer) and leaky ReLU activations.
- Leaky ReLU is a nonlinear function that give a non-zero output for a negative input. It aims to fix the "dying ReLU" problem and helps the gradients flow easier through the architecture.

### 20.2.6 Exercises

1. What will happen if we use standard ReLU activation rather than leaky ReLU?

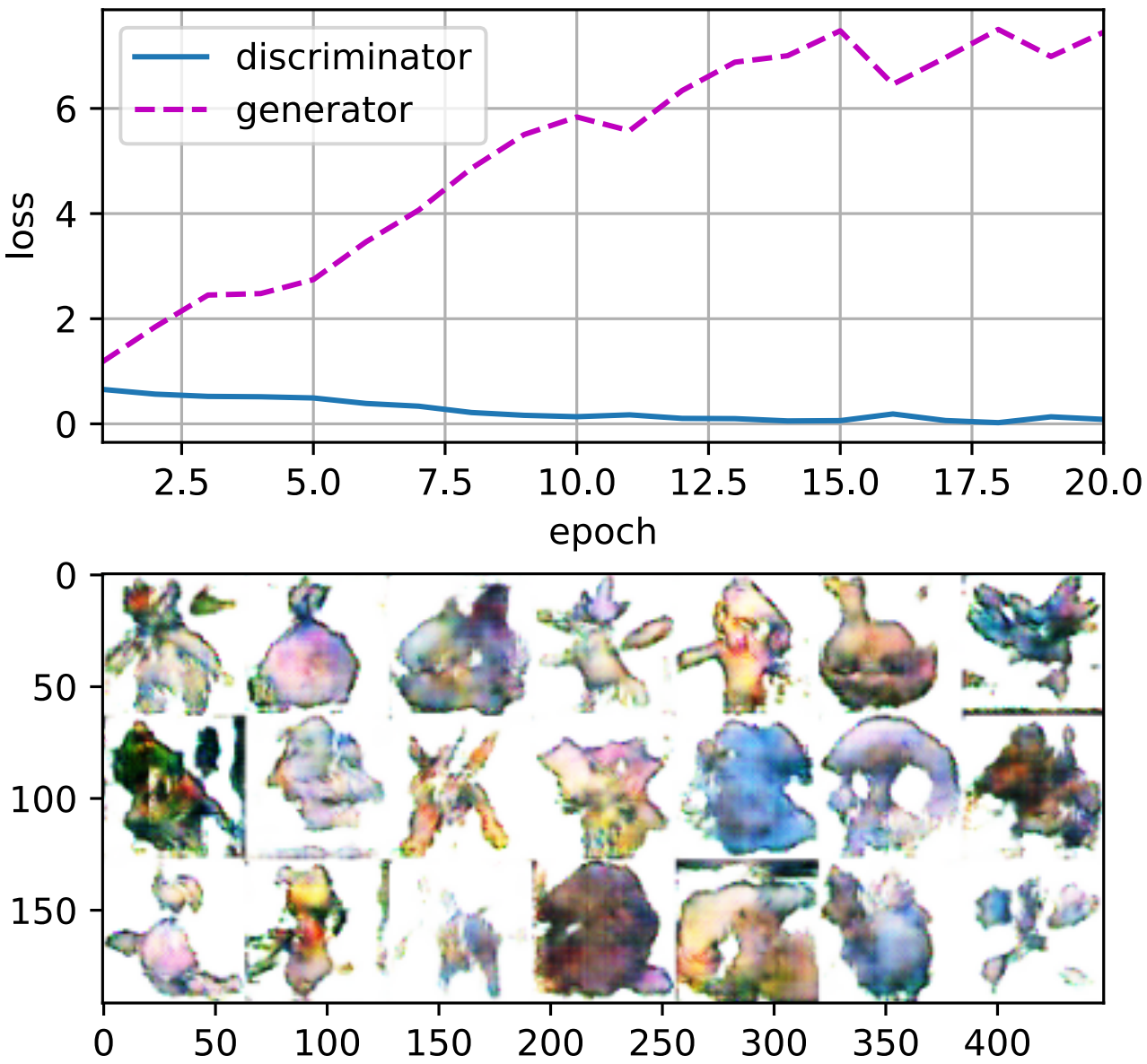

2. Apply DCGAN on Fashion-MNIST and see which category works well and which does not.

Discussions[277].

# A Mathematics for Deep Learning

**Brent Werness** (*Amazon*), **Rachel Hu** (*Amazon*), and authors of this book

One of the wonderful parts of modern deep learning is the fact that much of it can be understood and used without a full understanding of the mathematics below it. This is a sign that the field is maturing. Just as most software developers no longer need to worry about the theory of computable functions, neither should deep learning practitioners need to worry about the theoretical foundations of maximum likelihood learning.

But, we are not quite there yet.

In practice, you will sometimes need to understand how architectural choices influence gradient flow, or the implicit assumptions you make by training with a certain loss function. You might need to know what in the world entropy measures, and how it can help you understand exactly what bits-per-character means in your model. These all require deeper mathematical understanding.

This appendix aims to provide you the mathematical background you need to understand the core theory of modern deep learning, but it is not exhaustive. We will begin with examining linear algebra in greater depth. We develop a geometric understanding of all the common linear algebraic objects and operations that will enable us to visualize the effects of various transformations on our data. A key element is the development of the basics of eigen-decompositions.

We next develop the theory of differential calculus to the point that we can fully understand why the gradient is the direction of steepest descent, and why back-propagation takes the form it does. Integral calculus is then discussed to the degree needed to support our next topic, probability theory.

Problems encountered in practice frequently are not certain, and thus we need a language to speak about uncertain things. We review the theory of random variables and the most commonly encountered distributions so we may discuss models probabilistically. This provides the foundation for the naive Bayes classifier, a probabilistic classification technique.

Closely related to probability theory is the study of statistics. While statistics is far too large a field to do justice in a short section, we will introduce fundamental concepts that all machine learning practitioners should be aware of, in particular: evaluating and comparing estimators, conducting hypothesis tests, and constructing confidence intervals.

Last, we turn to the topic of information theory, which is the mathematical study of infor-mation storage and transmission. This provides the core language by which we may discuss quantitatively how much information a model holds on a domain of discourse.

Taken together, these form the core of the mathematical concepts needed to begin down the path towards a deep understanding of deep learning.

# A.1 Geometry and Linear Algebraic Operations

In Section 2.3, we encountered the basics of linear algebra and saw how it could be used to express common operations for transforming our data. Linear algebra is one of the key mathematical pillars underlying much of the work that we do in deep learning and in machine learning more broadly. While Section 2.3 contained enough machinery to communicate the mechanics of modern deep learning models, there is a lot more to the subject. In this section, we will go deeper, highlighting some geometric interpretations of linear algebra operations, and introducing a few fundamental concepts, including of eigenvalues and eigenvectors.

## A.1.1 Geometry of Vectors

First, we need to discuss the two common geometric interpretations of vectors, as either points or directions in space. Fundamentally, a vector is a list of numbers such as the Python list below.

```
v = [1, 7, 0, 1]
```

Mathematicians most often write this as either a *column* or *row* vector, which is to say either as

$$\mathbf{x} = \begin{bmatrix} 1 \\ 7 \\ 0 \\ 1 \end{bmatrix}, \tag{A.1}$$

or

$$\mathbf{x}^\top = \begin{bmatrix} 1 & 7 & 0 & 1 \end{bmatrix}. \tag{A.2}$$

These often have different interpretations, where data examples are column vectors and weights used to form weighted sums are row vectors. However, it can be beneficial to be flexible. As we have described in Section 2.3, though a single vector's default orientation is a column vector, for any matrix representing a tabular dataset, treating each data example as a row vector in the matrix is more conventional.

Given a vector, the first interpretation that we should give it is as a point in space. In two or three dimensions, we can visualize these points by using the components of the vectors to define the location of the points in space compared to a fixed reference called the *origin*. This can be seen in Fig. A.1.

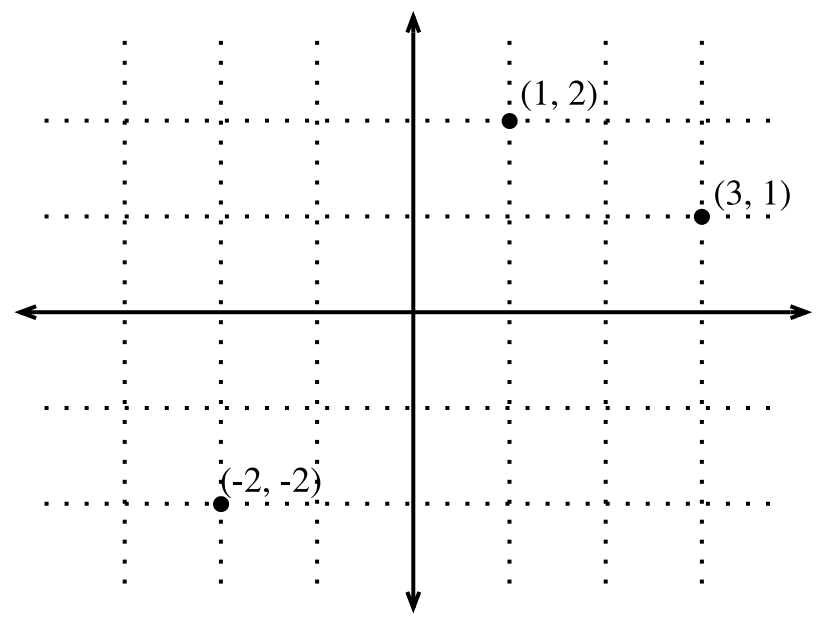

Fig. A.1 An illustration of visualizing vectors as points in the plane. The first component of the vector gives the $x$-coordinate, the second component gives the $y$-coordinate. Higher dimensions are analogous, although much harder to visualize.

This geometric point of view allows us to consider the problem on a more abstract level. No longer faced with some insurmountable seeming problem like classifying pictures as either cats or dogs, we can start considering tasks abstractly as collections of points in space and picturing the task as discovering how to separate two distinct clusters of points.

In parallel, there is a second point of view that people often take of vectors: as directions in space. Not only can we think of the vector $\mathbf{v} = [3, 2]^\top$ as the location 3 units to the right and 2 units up from the origin, we can also think of it as the direction itself to take 3 steps to the right and 2 steps up. In this way, we consider all the vectors in figure Fig. A.2 the same.

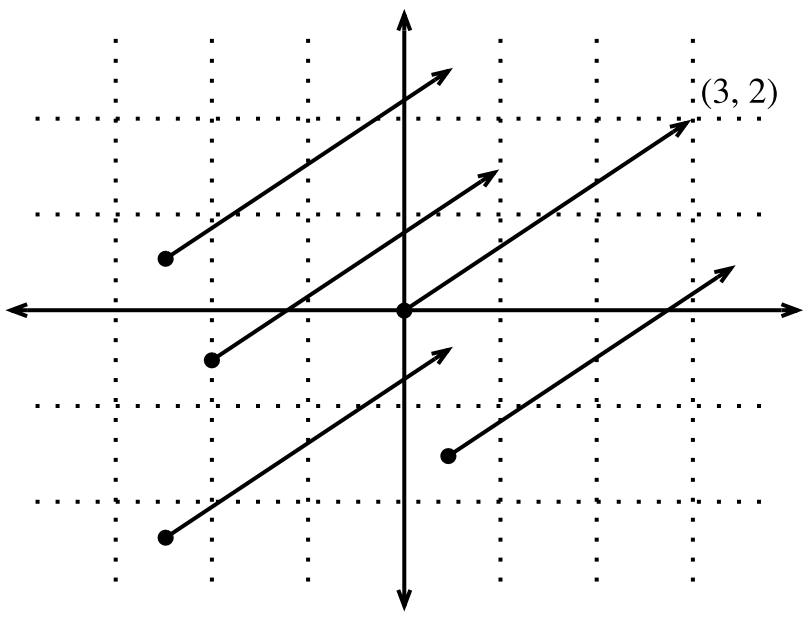

Fig. A.2 Any vector can be visualized as an arrow in the plane. In this case, every vector drawn is a representation of the vector $(3, 2)^\top$.

One of the benefits of this shift is that we can make visual sense of the act of vector addition. In particular, we follow the directions given by one vector, and then follow the directions given by the other, as is seen in Fig. A.3.

Vector subtraction has a similar interpretation. By considering the identity that $\mathbf{u} = \mathbf{v} + (\mathbf{u} -$

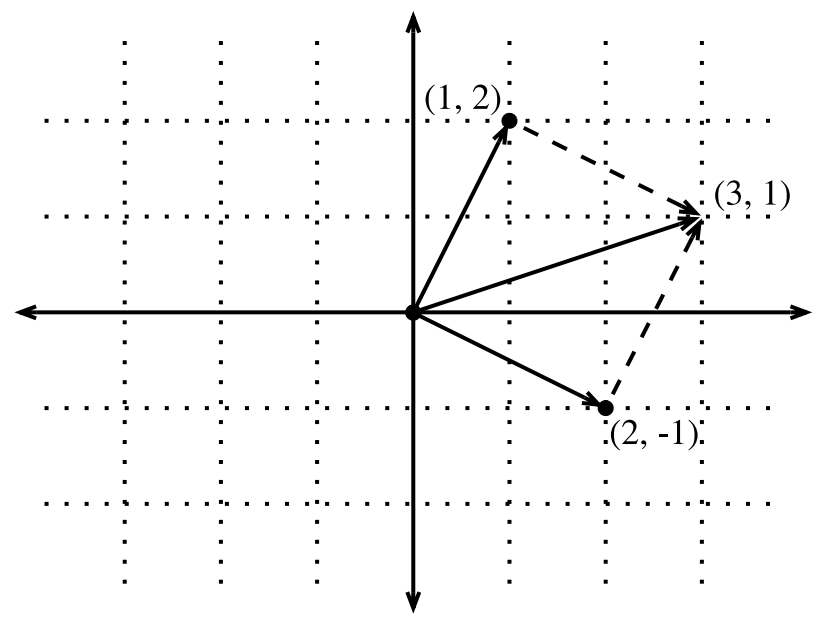

Fig. A.3 We can visualize vector addition by first following one vector, and then another.

$\mathbf{v}$), we see that the vector $\mathbf{u} - \mathbf{v}$ is the direction that takes us from the point $\mathbf{v}$ to the point $\mathbf{u}$.

## A.1.2 Dot Products and Angles

As we saw in Section 2.3, if we take two column vectors $\mathbf{u}$ and $\mathbf{v}$, we can form their dot product by computing:

$$\mathbf{u}^\top\mathbf{v} = \sum_i u_i \cdot v_i. \tag{A.3}$$

Because (A.3) is symmetric, we will mirror the notation of classical multiplication and write

$$\mathbf{u} \cdot \mathbf{v} = \mathbf{u}^\top\mathbf{v} = \mathbf{v}^\top\mathbf{u}, \tag{A.4}$$

to highlight the fact that exchanging the order of the vectors will yield the same answer.

The dot product (A.3) also admits a geometric interpretation: it is closely related to the angle between two vectors. Consider the angle shown in Fig. A.4.

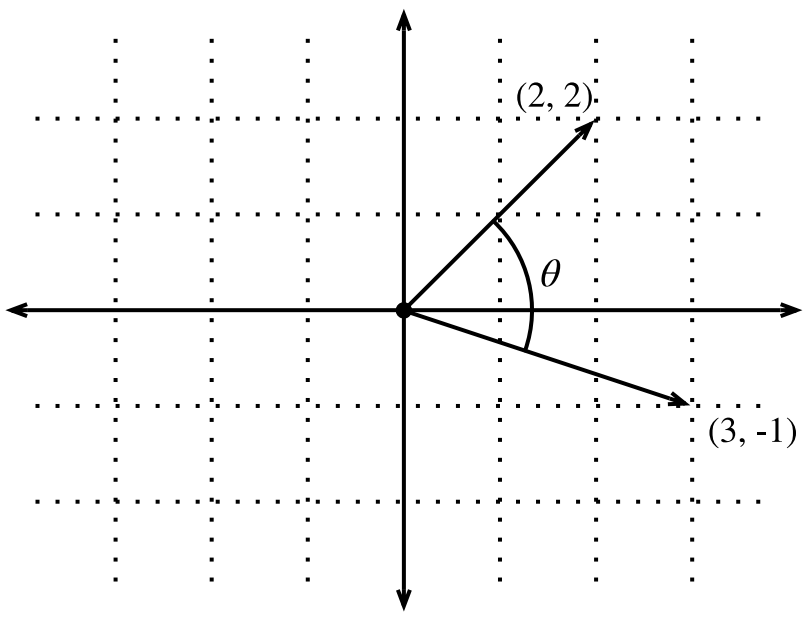

Fig. A.4 Between any two vectors in the plane there is a well defined angle $\theta$. We will see this angle is intimately tied to the dot product.

To start, let's consider two specific vectors:

$$\mathbf{v} = (r, 0) \text{ and } \mathbf{w} = (s\cos(\theta), s\sin(\theta)). \tag{A.5}$$

The vector $\mathbf{v}$ is length $r$ and runs parallel to the $x$-axis, and the vector $\mathbf{w}$ is of length $s$ and at angle $\theta$ with the $x$-axis. If we compute the dot product of these two vectors, we see that

$$\mathbf{v}\cdot\mathbf{w} = rs\cos(\theta) = \|\mathbf{v}\|\|\mathbf{w}\|\cos(\theta). \tag{A.6}$$

With some simple algebraic manipulation, we can rearrange terms to obtain

$$\theta = \arccos\left(\frac{\mathbf{v}\cdot\mathbf{w}}{\|\mathbf{v}\|\|\mathbf{w}\|}\right). \tag{A.7}$$

In short, for these two specific vectors, the dot product combined with the norms tell us the angle between the two vectors. This same fact is true in general. We will not derive the expression here, however, if we consider writing $\|\mathbf{v} - \mathbf{w}\|^2$ in two ways: one with the dot product, and the other geometrically using the law of cosines, we can obtain the full relationship. Indeed, for any two vectors $\mathbf{v}$ and $\mathbf{w}$, the angle between the two vectors is

$$\theta = \arccos\left(\frac{\mathbf{v}\cdot\mathbf{w}}{\|\mathbf{v}\|\|\mathbf{w}\|}\right). \tag{A.8}$$

This is a nice result since nothing in the computation references two-dimensions. Indeed, we can use this in three or three million dimensions without issue.

As a simple example, let's see how to compute the angle between a pair of vectors:

```
%matplotlib inline
import torch
import torchvision
from IPython import display
from torchvision import transforms
from d2l import torch as d2l

def angle(v, w):
    return torch.acos(v.dot(w) / (torch.norm(v) * torch.norm(w)))

angle(torch.tensor([0, 1, 2], dtype=torch.float32), torch.tensor([2.0, 3, 4]))
```

```
tensor(0.4190)
```

We will not use it right now, but it is useful to know that we will refer to vectors for which the angle is $\pi/2$ (or equivalently $90^\circ$) as being *orthogonal*. By examining the equation above, we see that this happens when $\theta = \pi/2$, which is the same thing as $\cos(\theta) = 0$. The only way this can happen is if the dot product itself is zero, and two vectors are orthogonal if and only if $\mathbf{v}\cdot\mathbf{w} = 0$. This will prove to be a helpful formula when understanding objects geometrically.

It is reasonable to ask: why is computing the angle useful? The answer comes in the kind of invariance we expect data to have. Consider an image, and a duplicate image, where every pixel value is the same but $10\%$ the brightness. The values of the individual pixels are in general far from the original values. Thus, if one computed the distance between the original image and the darker one, the distance can be large. However, for most ML applications, the *content* is the same—it is still an image of a cat as far as a cat/dog classifier is concerned. However, if we consider the angle, it is not hard to see that for any vector $\mathbf{v}$, the angle between $\mathbf{v}$ and $0.1\cdot\mathbf{v}$ is zero. This corresponds to the fact that scaling vectors keeps the same direction and just changes the length. The angle considers the darker image identical.

Examples like this are everywhere. In text, we might want the topic being discussed to not change if we write twice as long of document that says the same thing. For some encoding (such as counting the number of occurrences of words in some vocabulary), this corresponds to a doubling of the vector encoding the document, so again we can use the angle.

### Cosine Similarity

In ML contexts where the angle is employed to measure the closeness of two vectors, practitioners adopt the term *cosine similarity* to refer to the portion

$$\cos(\theta) = \frac{\mathbf{v}\cdot\mathbf{w}}{\|\mathbf{v}\|\|\mathbf{w}\|}. \tag{A.9}$$

The cosine takes a maximum value of $1$ when the two vectors point in the same direction, a minimum value of $-1$ when they point in opposite directions, and a value of $0$ when the two vectors are orthogonal. Note that if the components of high-dimensional vectors are sampled randomly with mean $0$, their cosine will nearly always be close to $0$.

## A.1.3 Hyperplanes

In addition to working with vectors, another key object that you must understand to go far in linear algebra is the *hyperplane*, a generalization to higher dimensions of a line (two dimensions) or of a plane (three dimensions). In an $d$-dimensional vector space, a hyperplane has $d-1$ dimensions and divides the space into two half-spaces.

Let's start with an example. Suppose that we have a column vector $\mathbf{w} = [2, 1]^\top$. We want to know, "what are the points $\mathbf{v}$ with $\mathbf{w}\cdot\mathbf{v} = 1$?" By recalling the connection between dot products and angles above (A.8), we can see that this is equivalent to

$$\|\mathbf{v}\|\|\mathbf{w}\|\cos(\theta) = 1 \quad\iff\quad \|\mathbf{v}\|\cos(\theta) = \frac{1}{\|\mathbf{w}\|} = \frac{1}{\sqrt{5}}. \tag{A.10}$$

If we consider the geometric meaning of this expression, we see that this is equivalent to saying that the length of the projection of $\mathbf{v}$ onto the direction of $\mathbf{w}$ is exactly $1/\|\mathbf{w}\|$, as is shown in Fig. A.5. The set of all points where this is true is a line at right angles to the vector $\mathbf{w}$. If we wanted, we could find the equation for this line and see that it is $2x + y = 1$ or equivalently $y = 1 - 2x$.

If we now look at what happens when we ask about the set of points with $\mathbf{w}\cdot\mathbf{v} > 1$ or $\mathbf{w}\cdot\mathbf{v} < 1$, we can see that these are cases where the projections are longer or shorter than

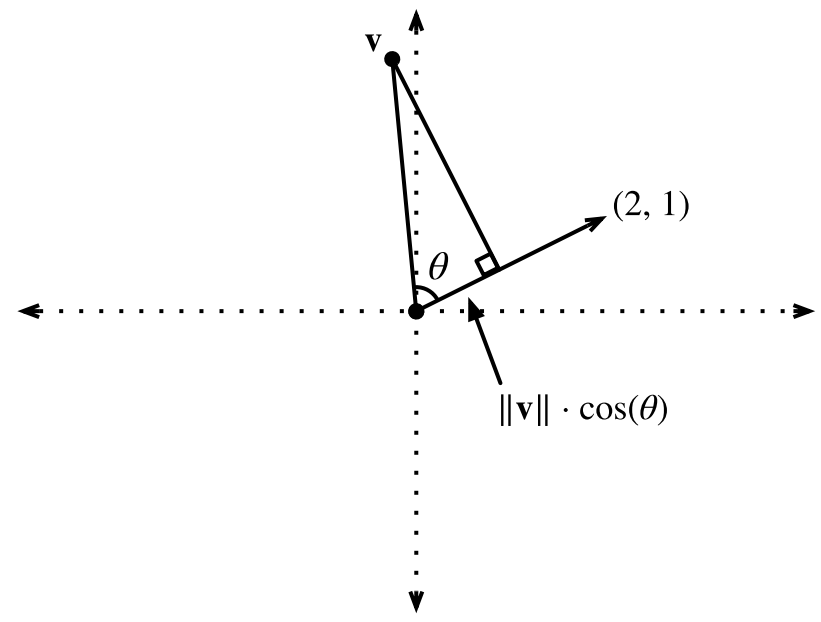

Fig. A.5 Recalling trigonometry, we see the formula $\|\mathbf{v}\|\cos(\theta)$ is the length of the projection of the vector $\mathbf{v}$ onto the direction of $\mathbf{w}$

$1/\|\mathbf{w}\|$, respectively. Thus, those two inequalities define either side of the line. In this way, we have found a way to cut our space into two halves, where all the points on one side have dot product below a threshold, and the other side above as we see in Fig. A.6.

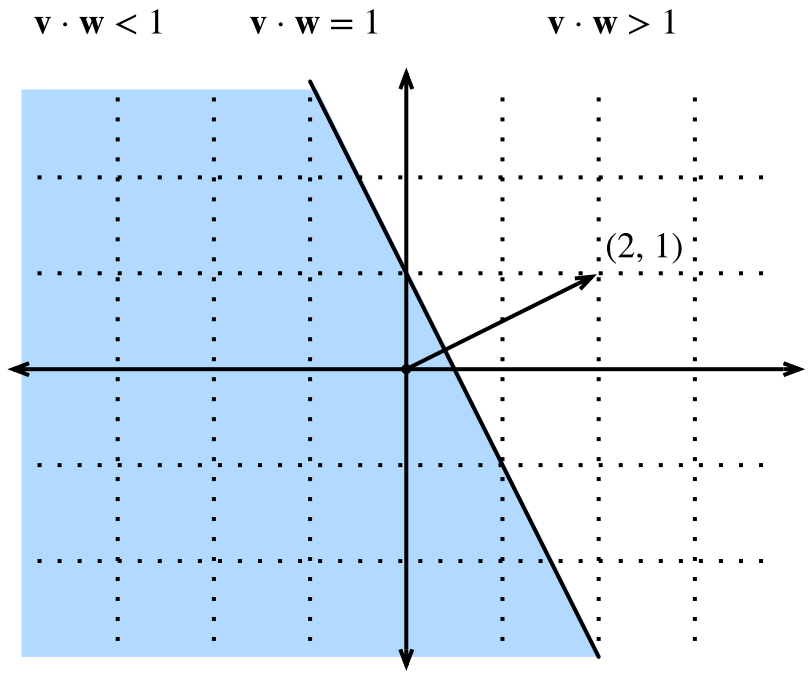

Fig. A.6 If we now consider the inequality version of the expression, we see that our hyperplane (in this case: just a line) separates the space into two halves.

The story in higher dimension is much the same. If we now take $\mathbf{w} = [1, 2, 3]^\top$ and ask about the points in three dimensions with $\mathbf{w} \cdot \mathbf{v} = 1$, we obtain a plane at right angles to the given vector $\mathbf{w}$. The two inequalities again define the two sides of the plane as is shown in Fig. A.7.

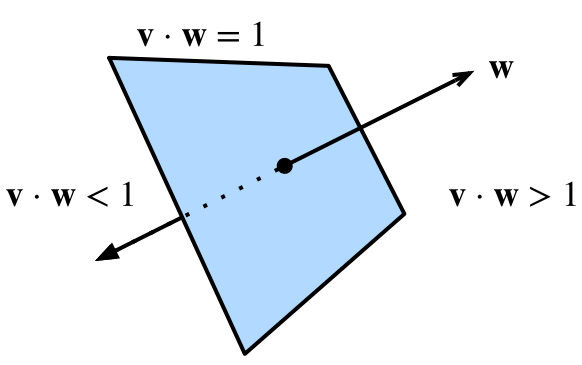

Fig. A.7 Hyperplanes in any dimension separate the space into two halves.

While our ability to visualize runs out at this point, nothing stops us from doing this in tens, hundreds, or billions of dimensions. This occurs often when thinking about machine learned models. For instance, we can understand linear classification models like those from Section 4.1, as methods to find hyperplanes that separate the different target classes. In this context, such hyperplanes are often referred to as *decision planes*. The majority of deep learned classification models end with a linear layer fed into a softmax, so one can interpret the role of the deep neural network to be to find a non-linear embedding such that the target classes can be separated cleanly by hyperplanes.

To give a hand-built example, notice that we can produce a reasonable model to classify tiny images of t-shirts and trousers from the Fashion-MNIST dataset (seen in Section 4.2) by just taking the vector between their means to define the decision plane and eyeball a crude threshold. First we will load the data and compute the averages.

```
# Load in the dataset
trans = []
trans.append(transforms.ToTensor())
trans = transforms.Compose(trans)
train = torchvision.datasets.FashionMNIST(root="../data", transform=trans,
                                          train=True, download=True)
test = torchvision.datasets.FashionMNIST(root="../data", transform=trans,
                                         train=False, download=True)

X_train_0 = torch.stack(
    [x[0] * 256 for x in train if x[1] == 0]).type(torch.float32)
X_train_1 = torch.stack(
    [x[0] * 256 for x in train if x[1] == 1]).type(torch.float32)
X_test = torch.stack(
    [x[0] * 256 for x in test if x[1] == 0 or x[1] == 1]).type(torch.float32)
y_test = torch.stack([torch.tensor(x[1]) for x in test
                      if x[1] == 0 or x[1] == 1]).type(torch.float32)

# Compute averages
ave_0 = torch.mean(X_train_0, axis=0)
ave_1 = torch.mean(X_train_1, axis=0)
```

It can be informative to examine these averages in detail, so let's plot what they look like. In this case, we see that the average indeed resembles a blurry image of a t-shirt.

```
# Plot average t-shirt
d2l.set_figsize()
d2l.plt.imshow(ave_0.reshape(28, 28).tolist(), cmap='Greys')
d2l.plt.show()
```

In the second case, we again see that the average resembles a blurry image of trousers.

```
# Plot average trousers
d2l.plt.imshow(ave_1.reshape(28, 28).tolist(), cmap='Greys')
d2l.plt.show()
```

In a fully machine learned solution, we would learn the threshold from the dataset. In this case, I simply eyeballed a threshold that looked good on the training data by hand.

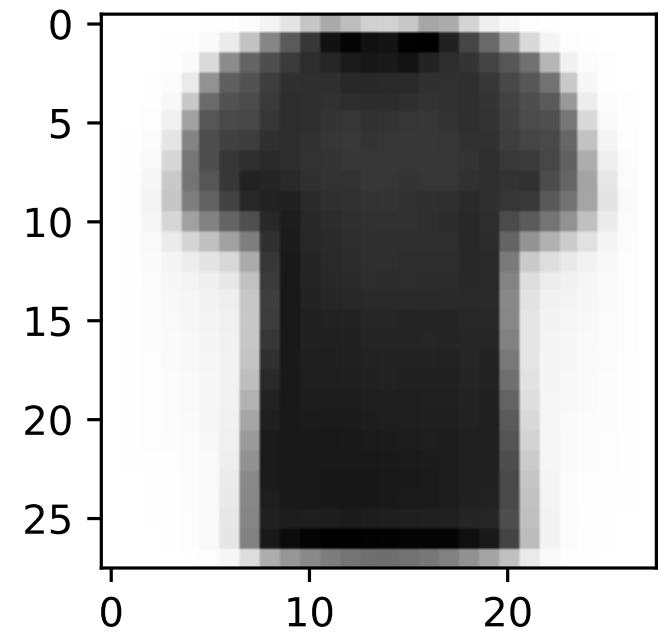

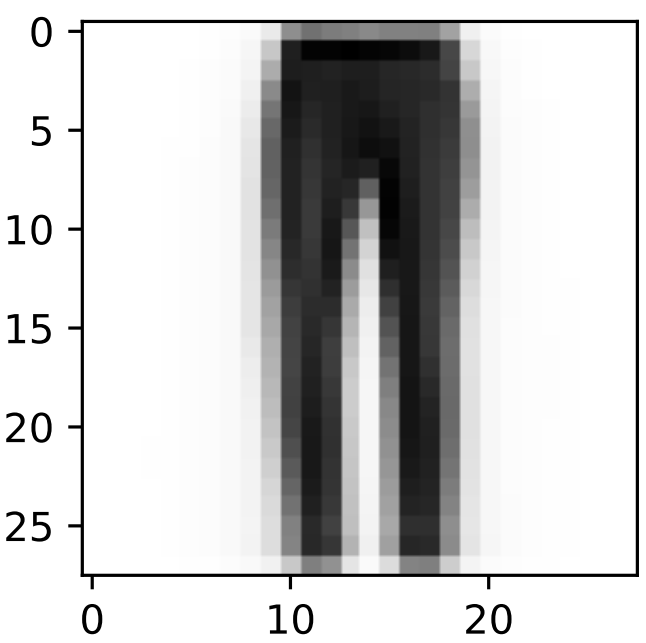

```
# Print test set accuracy with eyeballed threshold
w = (ave_1 - ave_0).T
# '@' is Matrix Multiplication operator in pytorch.
predictions = X_test.reshape(2000, -1) @ (w.flatten()) > -1500000

# Accuracy
torch.mean((predictions.type(y_test.dtype) == y_test).float(), dtype=torch.
↪float64)
```

```
tensor(0.7870, dtype=torch.float64)
```

### A.1.4 Geometry of Linear Transformations

Through Section 2.3 and the above discussions, we have a solid understanding of the geometry of vectors, lengths, and angles. However, there is one important object we have omitted discussing, and that is a geometric understanding of linear transformations represented by matrices. Fully internalizing what matrices can do to transform data between two potentially different high dimensional spaces takes significant practice, and is beyond the scope of this appendix. However, we can start building up intuition in two dimensions.

Suppose that we have some matrix:

$$\mathbf{A} = \begin{bmatrix} a & b \\ c & d \end{bmatrix}. \tag{A.11}$$

If we want to apply this to an arbitrary vector $\mathbf{v} = [x, y]^\top$, we multiply and see that

$$\begin{aligned} \mathbf{A}\mathbf{v} &= \begin{bmatrix} a & b \\ c & d \end{bmatrix} \begin{bmatrix} x \\ y \end{bmatrix} \\ &= \begin{bmatrix} ax + by \\ cx + dy \end{bmatrix} \\ &= x \begin{bmatrix} a \\ c \end{bmatrix} + y \begin{bmatrix} b \\ d \end{bmatrix} \\ &= x \left\{ \mathbf{A} \begin{bmatrix} 1 \\ 0 \end{bmatrix} \right\} + y \left\{ \mathbf{A} \begin{bmatrix} 0 \\ 1 \end{bmatrix} \right\}. \end{aligned} \tag{A.12}$$

This may seem like an odd computation, where something clear became somewhat impenetrable. However, it tells us that we can write the way that a matrix transforms *any* vector in terms of how it transforms *two specific vectors*: $[1, 0]^\top$ and $[0, 1]^\top$. This is worth considering for a moment. We have essentially reduced an infinite problem (what happens to any pair of real numbers) to a finite one (what happens to these specific vectors). These vectors are an example a *basis*, where we can write any vector in our space as a weighted sum of these *basis vectors*.

Let's draw what happens when we use the specific matrix

$$\mathbf{A} = \begin{bmatrix} 1 & 2 \\ -1 & 3 \end{bmatrix}. \tag{A.13}$$

If we look at the specific vector $\mathbf{v} = [2, -1]^\top$, we see this is $2 \cdot [1, 0]^\top + -1 \cdot [0, 1]^\top$, and thus we know that the matrix $A$ will send this to $2(\mathbf{A}[1, 0]^\top) + -1(\mathbf{A}[0, 1])^\top = 2[1, -1]^\top - [2, 3]^\top = [0, -5]^\top$. If we follow this logic through carefully, say by considering the grid of all integer pairs of points, we see that what happens is that the matrix multiplication can skew, rotate, and scale the grid, but the grid structure must remain as you see in Fig. A.8.

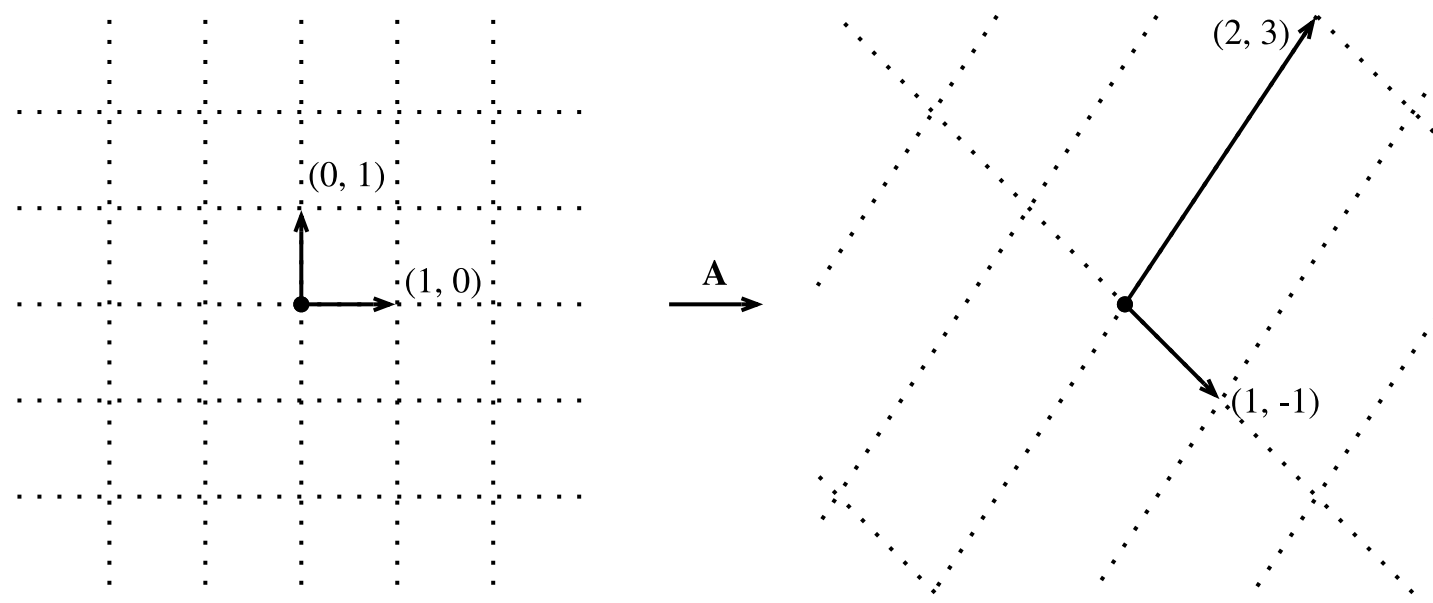

Fig. A.8 The matrix $\mathbf{A}$ acting on the given basis vectors. Notice how the entire grid is transported along with it.

This is the most important intuitive point to internalize about linear transformations represented by matrices. Matrices are incapable of distorting some parts of space differently than others. All they can do is take the original coordinates on our space and skew, rotate, and scale them.

Some distortions can be severe. For instance the matrix

$$\mathbf{B} = \begin{bmatrix} 2 & -1 \\ 4 & -2 \end{bmatrix}, \tag{A.14}$$

compresses the entire two-dimensional plane down to a single line. Identifying and working with such transformations are the topic of a later section, but geometrically we can see that this is fundamentally different from the types of transformations we saw above. For instance, the result from matrix $\mathbf{A}$ can be "bent back" to the original grid. The results from matrix $\mathbf{B}$ cannot because we will never know where the vector $[1, 2]^\top$ came from—was it $[1, 1]^\top$ or $[0, -1]^\top$?

While this picture was for a $2 \times 2$ matrix, nothing prevents us from taking the lessons learned into higher dimensions. If we take similar basis vectors like $[1, 0, \ldots, 0]$ and see where our matrix sends them, we can start to get a feeling for how the matrix multiplication distorts the entire space in whatever dimension space we are dealing with.

### A.1.5 Linear Dependence

Consider again the matrix

$$\mathbf{B} = \begin{bmatrix} 2 & -1 \\ 4 & -2 \end{bmatrix}. \tag{A.15}$$

This compresses the entire plane down to live on the single line $y = 2x$. The question now arises: is there some way we can detect this just looking at the matrix itself? The answer is that indeed we can. Let's take $\mathbf{b}_1 = [2, 4]^\top$ and $\mathbf{b}_2 = [-1, -2]^\top$ be the two columns of $\mathbf{B}$. Remember that we can write everything transformed by the matrix $\mathbf{B}$ as a weighted sum of the columns of the matrix: like $a_1\mathbf{b}_1 + a_2\mathbf{b}_2$. We call this a *linear combination*. The fact that $\mathbf{b}_1 = -2 \cdot \mathbf{b}_2$ means that we can write any linear combination of those two columns entirely in terms of say $\mathbf{b}_2$ since

$$a_1\mathbf{b}_1 + a_2\mathbf{b}_2 = -2a_1\mathbf{b}_2 + a_2\mathbf{b}_2 = (a_2 - 2a_1)\mathbf{b}_2. \tag{A.16}$$

This means that one of the columns is, in a sense, redundant because it does not define a unique direction in space. This should not surprise us too much since we already saw that this matrix collapses the entire plane down into a single line. Moreover, we see that the linear dependence $\mathbf{b}_1 = -2 \cdot \mathbf{b}_2$ captures this. To make this more symmetrical between the two vectors, we will write this as

$$\mathbf{b}_1 + 2 \cdot \mathbf{b}_2 = 0. \tag{A.17}$$

In general, we will say that a collection of vectors $\mathbf{v}_1, \ldots, \mathbf{v}_k$ are *linearly dependent* if there exist coefficients $a_1, \ldots, a_k$ *not all equal to zero* so that

$$\sum_{i=1}^{k} a_i \mathbf{v_i} = 0. \tag{A.18}$$

In this case, we can solve for one of the vectors in terms of some combination of the others, and effectively render it redundant. Thus, a linear dependence in the columns of a matrix is a witness to the fact that our matrix is compressing the space down to some lower dimension. If there is no linear dependence we say the vectors are *linearly independent*. If the columns of a matrix are linearly independent, no compression occurs and the operation can be undone.

## A.1.6 Rank

If we have a general $n \times m$ matrix, it is reasonable to ask what dimension space the matrix maps into. A concept known as the *rank* will be our answer. In the previous section, we noted that a linear dependence bears witness to compression of space into a lower dimension and so we will be able to use this to define the notion of rank. In particular, the rank of a matrix $\mathbf{A}$ is the largest number of linearly independent columns amongst all subsets of columns. For example, the matrix

$$\mathbf{B} = \begin{bmatrix} 2 & 4 \\ -1 & -2 \end{bmatrix}, \tag{A.19}$$

has $\operatorname{rank}(B) = 1$, since the two columns are linearly dependent, but either column by itself is not linearly dependent. For a more challenging example, we can consider

$$\mathbf{C} = \begin{bmatrix} 1 & 3 & 0 & -1 & 0 \\ -1 & 0 & 1 & 1 & -1 \\ 0 & 3 & 1 & 0 & -1 \\ 2 & 3 & -1 & -2 & 1 \end{bmatrix}, \tag{A.20}$$

and show that $\mathbf{C}$ has rank two since, for instance, the first two columns are linearly independent, however any of the four collections of three columns are dependent.

This procedure, as described, is very inefficient. It requires looking at every subset of the columns of our given matrix, and thus is potentially exponential in the number of columns. Later we will see a more computationally efficient way to compute the rank of a matrix, but for now, this is sufficient to see that the concept is well defined and understand the meaning.

## A.1.7 Invertibility

We have seen above that multiplication by a matrix with linearly dependent columns cannot be undone, i.e., there is no inverse operation that can always recover the input. However, multiplication by a full-rank matrix (i.e., some $\mathbf{A}$ that is $n \times n$ matrix with rank $n$), we should always be able to undo it. Consider the matrix

$$\mathbf{I} = \begin{bmatrix} 1 & 0 & \cdots & 0 \\ 0 & 1 & \cdots & 0 \\ \vdots & \vdots & \ddots & \vdots \\ 0 & 0 & \cdots & 1 \end{bmatrix}. \tag{A.21}$$

which is the matrix with ones along the diagonal, and zeros elsewhere. We call this the *identity* matrix. It is the matrix which leaves our data unchanged when applied. To find a matrix which undoes what our matrix $\mathbf{A}$ has done, we want to find a matrix $\mathbf{A}^{-1}$ such that

$$\mathbf{A}^{-1}\mathbf{A} = \mathbf{A}\mathbf{A}^{-1} = \mathbf{I}. \tag{A.22}$$

If we look at this as a system, we have $n \times n$ unknowns (the entries of $\mathbf{A}^{-1}$) and $n \times n$ equations (the equality that needs to hold between every entry of the product $\mathbf{A}^{-1}\mathbf{A}$ and every entry of $\mathbf{I}$) so we should generically expect a solution to exist. Indeed, in the next section we will see a quantity called the *determinant*, which has the property that as long as the determinant is not zero, we can find a solution. We call such a matrix $\mathbf{A}^{-1}$ the *inverse* matrix. As an example, if $\mathbf{A}$ is the general $2 \times 2$ matrix

$$\mathbf{A} = \begin{bmatrix} a & b \\ c & d \end{bmatrix}, \tag{A.23}$$

then we can see that the inverse is

$$\frac{1}{ad - bc} \begin{bmatrix} d & -b \\ -c & a \end{bmatrix}. \tag{A.24}$$

We can test to see this by seeing that multiplying by the inverse given by the formula above works in practice.

```
M = torch.tensor([[1, 2], [1, 4]], dtype=torch.float32)
M_inv = torch.tensor([[2, -1], [-0.5, 0.5]])
M_inv @ M
```

```
tensor([[1., 0.],
        [0., 1.]])
```

## Numerical Issues

While the inverse of a matrix is useful in theory, we must say that most of the time we do not wish to *use* the matrix inverse to solve a problem in practice. In general, there are far more numerically stable algorithms for solving linear equations like

$$\mathbf{A}\mathbf{x} = \mathbf{b}, \tag{A.25}$$

than computing the inverse and multiplying to get

$$\mathbf{x} = \mathbf{A}^{-1}\mathbf{b}. \tag{A.26}$$

Just as division by a small number can lead to numerical instability, so can inversion of a matrix which is close to having low rank.

Moreover, it is common that the matrix $\mathbf{A}$ is *sparse*, which is to say that it contains only a small number of non-zero values. If we were to explore examples, we would see that this does not mean the inverse is sparse. Even if $\mathbf{A}$ was a $1$ million by $1$ million matrix with only $5$ million non-zero entries (and thus we need only store those $5$ million), the inverse will typically have almost every entry non-negative, requiring us to store all $1\text{M}^2$ entries—that is $1$ trillion entries!

While we do not have time to dive all the way into the thorny numerical issues frequently encountered when working with linear algebra, we want to provide you with some intuition about when to proceed with caution, and generally avoiding inversion in practice is a good rule of thumb.

## A.1.8 Determinant

The geometric view of linear algebra gives an intuitive way to interpret a fundamental quantity known as the *determinant*. Consider the grid image from before, but now with a highlighted region (Fig. A.9).

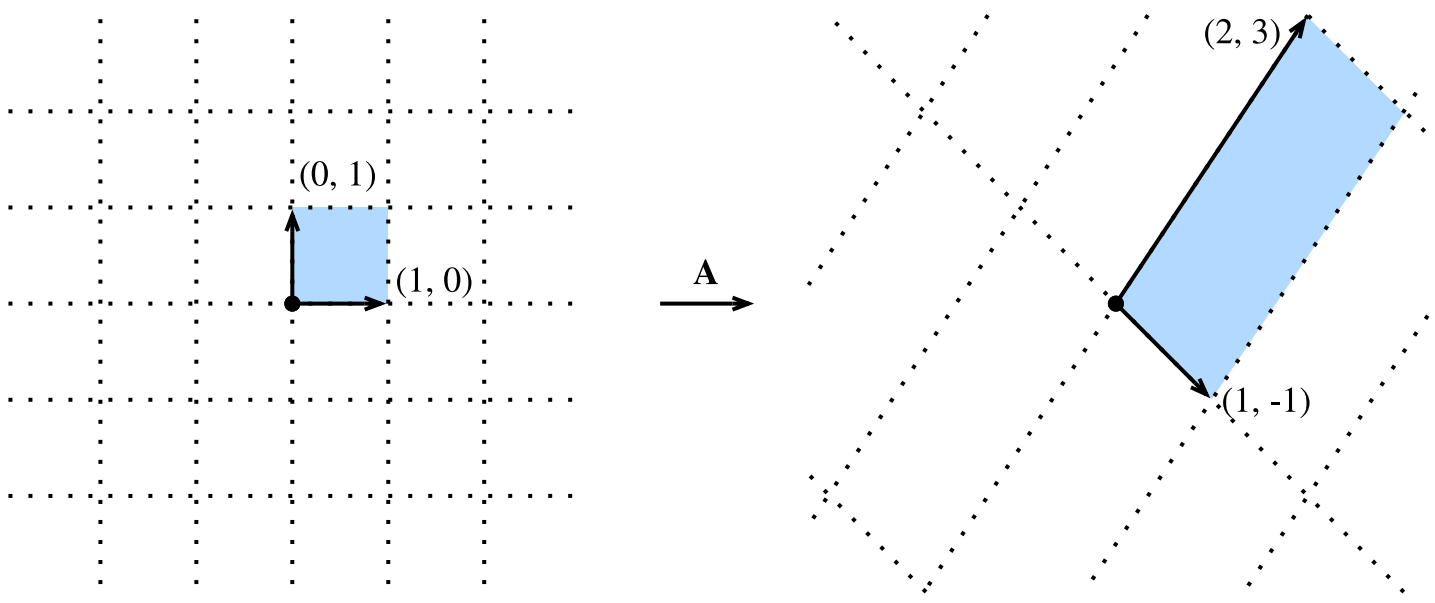

Fig. A.9 The matrix $\mathbf{A}$ again distorting the grid. This time, I want to draw particular attention to what happens to the highlighted square.

Look at the highlighted square. This is a square with edges given by $(0, 1)$ and $(1, 0)$ and thus it has area one. After $\mathbf{A}$ transforms this square, we see that it becomes a parallelogram. There is no reason this parallelogram should have the same area that we started with, and indeed in the specific case shown here of

$$\mathbf{A} = \begin{bmatrix} 1 & 2 \\ -1 & 3 \end{bmatrix}, \tag{A.27}$$

it is an exercise in coordinate geometry to compute the area of this parallelogram and obtain that the area is $5$.

In general, if we have a matrix

$$\mathbf{A} = \begin{bmatrix}a & b \\ c & d\end{bmatrix}, \tag{A.28}$$

we can see with some computation that the area of the resulting parallelogram is $ad - bc$. This area is referred to as the *determinant*.

Let's check this quickly with some example code.

```
torch.det(torch.tensor([[1, -1], [2, 3]], dtype=torch.float32))
```

```
tensor(5.)
```

The eagle-eyed amongst us will notice that this expression can be zero or even negative. For the negative term, this is a matter of convention taken generally in mathematics: if the matrix flips the figure, we say the area is negated. Let's see now that when the determinant is zero, we learn more.

Let's consider

$$\mathbf{B} = \begin{bmatrix} 2 & 4 \\ -1 & -2 \end{bmatrix}. \tag{A.29}$$

If we compute the determinant of this matrix, we get $2 \cdot (-2) - 4 \cdot (-1) = 0$. Given our understanding above, this makes sense. $\mathbf{B}$ compresses the square from the original image down to a line segment, which has zero area. And indeed, being compressed into a lower dimensional space is the only way to have zero area after the transformation. Thus we see the following result is true: a matrix $A$ is invertible if and only if the determinant is not equal to zero.

As a final comment, imagine that we have any figure drawn on the plane. Thinking like computer scientists, we can decompose that figure into a collection of little squares so that the area of the figure is in essence just the number of squares in the decomposition. If we now transform that figure by a matrix, we send each of these squares to parallelograms, each one of which has area given by the determinant. We see that for any figure, the determinant gives the (signed) number that a matrix scales the area of any figure.

Computing determinants for larger matrices can be laborious, but the intuition is the same. The determinant remains the factor that $n \times n$ matrices scale $n$-dimensional volumes.

### A.1.9 Tensors and Common Linear Algebra Operations

In Section 2.3 the concept of tensors was introduced. In this section, we will dive more deeply into tensor contractions (the tensor equivalent of matrix multiplication), and see how it can provide a unified view on a number of matrix and vector operations.

With matrices and vectors we knew how to multiply them to transform data. We need to have a similar definition for tensors if they are to be useful to us. Think about matrix multiplication:

$$\mathbf{C} = \mathbf{A}\mathbf{B}, \tag{A.30}$$

or equivalently

$$c_{i,j} = \sum_k a_{i,k} b_{k,j}. \tag{A.31}$$

This pattern is one we can repeat for tensors. For tensors, there is no one case of what to sum over that can be universally chosen, so we need specify exactly which indices we want to sum over. For instance we could consider

$$y_{il} = \sum_{jk} x_{ijkl} a_{jk}. \tag{A.32}$$

Such a transformation is called a *tensor contraction*. It can represent a far more flexible family of transformations that matrix multiplication alone.

As a often-used notational simplification, we can notice that the sum is over exactly those indices that occur more than once in the expression, thus people often work with *Einstein notation*, where the summation is implicitly taken over all repeated indices. This gives the compact expression:

$$y_{il} = x_{ijkl} a_{jk}. \tag{A.33}$$

## Common Examples from Linear Algebra

Let's see how many of the linear algebraic definitions we have seen before can be expressed in this compressed tensor notation:

- $\mathbf{v} \cdot \mathbf{w} = \sum_i v_i w_i$
- $\|\mathbf{v}\|_2^2 = \sum_i v_i v_i$
- $(\mathbf{A}\mathbf{v})_i = \sum_j a_{ij} v_j$
- $(\mathbf{A}\mathbf{B})_{ik} = \sum_j a_{ij} b_{jk}$
- $\mathrm{tr}(\mathbf{A}) = \sum_i a_{ii}$

In this way, we can replace a myriad of specialized notations with short tensor expressions.

## Expressing in Code

Tensors may flexibly be operated on in code as well. As seen in Section 2.3, we can create tensors as is shown below.

```
# Define tensors
B = torch.tensor([[[1, 2, 3], [4, 5, 6]], [[7, 8, 9], [10, 11, 12]]])
A = torch.tensor([[1, 2], [3, 4]])
v = torch.tensor([1, 2])

# Print out the shapes
A.shape, B.shape, v.shape
```

```
(torch.Size([2, 2]), torch.Size([2, 2, 3]), torch.Size([2]))
```

Einstein summation has been implemented directly. The indices that occurs in the Einstein summation can be passed as a string, followed by the tensors that are being acted upon. For instance, to implement matrix multiplication, we can consider the Einstein summation seen above ($\mathbf{A}\mathbf{v} = a_{ij}v_j$) and strip out the indices themselves to get the implementation:

```
# Reimplement matrix multiplication
torch.einsum("ij, j -> i", A, v), A@v
```

```
(tensor([ 5, 11]), tensor([ 5, 11]))
```

This is a highly flexible notation. For instance if we want to compute what would be traditionally written as

$$c_{kl} = \sum_{ij} \mathbf{b}_{ijk}\mathbf{a}_{il}v_j. \tag{A.34}$$

it can be implemented via Einstein summation as:

```
torch.einsum("ijk, il, j -> kl", B, A, v)
```

```
tensor([[ 90, 126],
        [102, 144],
        [114, 162]])
```

This notation is readable and efficient for humans, however bulky if for whatever reason we need to generate a tensor contraction programmatically. For this reason, `einsum` provides an alternative notation by providing integer indices for each tensor. For example, the same tensor contraction can also be written as:

```
# PyTorch does not support this type of notation.
```

Either notation allows for concise and efficient representation of tensor contractions in code.

### A.1.10 Summary

- Vectors can be interpreted geometrically as either points or directions in space.

- Dot products define the notion of angle to arbitrarily high-dimensional spaces.
- Hyperplanes are high-dimensional generalizations of lines and planes. They can be used to define decision planes that are often used as the last step in a classification task.
- Matrix multiplication can be geometrically interpreted as uniform distortions of the underlying coordinates. They represent a very restricted, but mathematically clean, way to transform vectors.
- Linear dependence is a way to tell when a collection of vectors are in a lower dimensional space than we would expect (say you have $3$ vectors living in a $2$-dimensional space). The rank of a matrix is the size of the largest subset of its columns that are linearly independent.
- When a matrix's inverse is defined, matrix inversion allows us to find another matrix that undoes the action of the first. Matrix inversion is useful in theory, but requires care in practice owing to numerical instability.
- Determinants allow us to measure how much a matrix expands or contracts a space. A nonzero determinant implies an invertible (non-singular) matrix and a zero-valued determinant means that the matrix is non-invertible (singular).
- Tensor contractions and Einstein summation provide for a neat and clean notation for expressing many of the computations that are seen in machine learning.

## A.1.11 Exercises

1. What is the angle between

$$\vec{v}_1 = \begin{bmatrix} 1 \\ 0 \\ -1 \\ 2 \end{bmatrix}, \qquad \vec{v}_2 = \begin{bmatrix} 3 \\ 1 \\ 0 \\ 1 \end{bmatrix}? \tag{A.35}$$

2. True or false: $\begin{bmatrix} 1 & 2 \\ 0 & 1 \end{bmatrix}$ and $\begin{bmatrix} 1 & -2 \\ 0 & 1 \end{bmatrix}$ are inverses of one another?

3. Suppose that we draw a shape in the plane with area $100\mathrm{m}^2$. What is the area after transforming the figure by the matrix

$$\begin{bmatrix} 2 & 3 \\ 1 & 2 \end{bmatrix}. \tag{A.36}$$

4. Which of the following sets of vectors are linearly independent?

- $\left\{ \begin{pmatrix} 1 \\ 0 \\ -1 \end{pmatrix}, \begin{pmatrix} 2 \\ 1 \\ -1 \end{pmatrix}, \begin{pmatrix} 3 \\ 1 \\ 1 \end{pmatrix} \right\}$

- $\left\{\begin{pmatrix}3\\1\\1\end{pmatrix},\begin{pmatrix}1\\1\\1\end{pmatrix},\begin{pmatrix}0\\0\\0\end{pmatrix}\right\}$
- $\left\{\begin{pmatrix}1\\1\\0\end{pmatrix},\begin{pmatrix}0\\1\\-1\end{pmatrix},\begin{pmatrix}1\\0\\1\end{pmatrix}\right\}$

5. Suppose that you have a matrix written as $A = \begin{bmatrix}c\\d\end{bmatrix} \cdot \begin{bmatrix}a & b\end{bmatrix}$ for some choice of values $a, b, c$, and $d$. True or false: the determinant of such a matrix is always 0?
6. The vectors $e_1 = \begin{bmatrix}1\\0\end{bmatrix}$ and $e_2 = \begin{bmatrix}0\\1\end{bmatrix}$ are orthogonal. What is the condition on a matrix $A$ so that $Ae_1$ and $Ae_2$ are orthogonal?
7. How can you write $\mathrm{tr}(\mathbf{A}^4)$ in Einstein notation for an arbitrary matrix $A$?

Discussions[278].

## A.2 Eigendecompositions

Eigenvalues are often one of the most useful notions we will encounter when studying linear algebra, however, as a beginner, it is easy to overlook their importance. Below, we introduce eigendecomposition and try to convey some sense of just why it is so important.

Suppose that we have a matrix $A$ with the following entries:

$$\mathbf{A} = \begin{bmatrix}2 & 0\\0 & -1\end{bmatrix}. \tag{A.1}$$

If we apply $A$ to any vector $\mathbf{v} = [x, y]^\top$, we obtain a vector $\mathbf{A}\mathbf{v} = [2x, -y]^\top$. This has an intuitive interpretation: stretch the vector to be twice as wide in the $x$-direction, and then flip it in the $y$-direction.

However, there are *some* vectors for which something remains unchanged. Namely $[1, 0]^\top$ gets sent to $[2, 0]^\top$ and $[0, 1]^\top$ gets sent to $[0, -1]^\top$. These vectors are still in the same line, and the only modification is that the matrix stretches them by a factor of 2 and $-1$ respectively. We call such vectors *eigenvectors* and the factor they are stretched by *eigenvalues*.

In general, if we can find a number $\lambda$ and a vector $\mathbf{v}$ such that

$$\mathbf{A}\mathbf{v} = \lambda \mathbf{v}. \tag{A.2}$$

We say that $\mathbf{v}$ is an eigenvector for $A$ and $\lambda$ is an eigenvalue.

### A.2.1 Finding Eigenvalues

Let's figure out how to find them. By subtracting off the $\lambda \mathbf{v}$ from both sides, and then factoring out the vector, we see the above is equivalent to:

$$(\mathbf{A} - \lambda \mathbf{I})\mathbf{v} = 0. \tag{A.3}$$

For (A.3) to happen, we see that $(\mathbf{A} - \lambda \mathbf{I})$ must compress some direction down to zero, hence it is not invertible, and thus the determinant is zero. Thus, we can find the *eigenvalues* by finding for what $\lambda$ is $\det(\mathbf{A} - \lambda \mathbf{I}) = 0$. Once we find the eigenvalues, we can solve $\mathbf{A}\mathbf{v} = \lambda \mathbf{v}$ to find the associated *eigenvector(s)*.

#### An Example

Let's see this with a more challenging matrix

$$\mathbf{A} = \begin{bmatrix} 2 & 1 \\ 2 & 3 \end{bmatrix}. \tag{A.4}$$

If we consider $\det(\mathbf{A} - \lambda \mathbf{I}) = 0$, we see this is equivalent to the polynomial equation $0 = (2-\lambda)(3-\lambda)-2 = (4-\lambda)(1-\lambda)$. Thus, two eigenvalues are $4$ and $1$. To find the associated vectors, we then need to solve

$$\begin{bmatrix} 2 & 1 \\ 2 & 3 \end{bmatrix}\begin{bmatrix} x \\ y \end{bmatrix} = \begin{bmatrix} x \\ y \end{bmatrix} \quad \text{and} \quad \begin{bmatrix} 2 & 1 \\ 2 & 3 \end{bmatrix}\begin{bmatrix} x \\ y \end{bmatrix} = \begin{bmatrix} 4x \\ 4y \end{bmatrix}. \tag{A.5}$$

We can solve this with the vectors $[1, -1]^\top$ and $[1, 2]^\top$ respectively.

We can check this in code using the built-in `numpy.linalg.eig` routine.

```
%matplotlib inline
import torch
from IPython import display
from d2l import torch as d2l

torch.linalg.eig(torch.tensor([[2, 1], [2, 3]], dtype=torch.float64))
```

```
torch.return_types.linalg_eig(
eigenvalues=tensor([1.+0.j, 4.+0.j], dtype=torch.complex128),
eigenvectors=tensor([[-0.7071+0.j, -0.4472+0.j],
        [ 0.7071+0.j, -0.8944+0.j]], dtype=torch.complex128))
```

Note that `numpy` normalizes the eigenvectors to be of length one, whereas we took ours to be of arbitrary length. Additionally, the choice of sign is arbitrary. However, the vectors computed are parallel to the ones we found by hand with the same eigenvalues.

### A.2.2 Decomposing Matrices

Let's continue the previous example one step further. Let

$$\mathbf{W} = \begin{bmatrix} 1 & 1 \\ -1 & 2 \end{bmatrix}, \tag{A.6}$$

be the matrix where the columns are the eigenvectors of the matrix $\mathbf{A}$. Let

$$\boldsymbol{\Sigma} = \begin{bmatrix} 1 & 0 \\ 0 & 4 \end{bmatrix}, \tag{A.7}$$

be the matrix with the associated eigenvalues on the diagonal. Then the definition of eigenvalues and eigenvectors tells us that

$$\mathbf{A}\mathbf{W} = \mathbf{W}\boldsymbol{\Sigma}. \tag{A.8}$$

The matrix $W$ is invertible, so we may multiply both sides by $W^{-1}$ on the right, we see that we may write

$$\mathbf{A} = \mathbf{W}\boldsymbol{\Sigma}\mathbf{W}^{-1}. \tag{A.9}$$

In the next section we will see some nice consequences of this, but for now we need only know that such a decomposition will exist as long as we can find a full collection of linearly independent eigenvectors (so that $W$ is invertible).

### A.2.3 Operations on Eigendecompositions

One nice thing about eigendecompositions (A.9) is that we can write many operations we usually encounter cleanly in terms of the eigendecomposition. As a first example, consider:

$$\mathbf{A}^n = \overbrace{\mathbf{A} \cdots \mathbf{A}}^{n \text{ times}} = \overbrace{(\mathbf{W}\boldsymbol{\Sigma}\mathbf{W}^{-1}) \cdots (\mathbf{W}\boldsymbol{\Sigma}\mathbf{W}^{-1})}^{n \text{ times}} = \mathbf{W}\overbrace{\boldsymbol{\Sigma} \cdots \boldsymbol{\Sigma}}^{n \text{ times}}\mathbf{W}^{-1} = \mathbf{W}\boldsymbol{\Sigma}^n\mathbf{W}^{-1}. \tag{A.10}$$

This tells us that for any positive power of a matrix, the eigendecomposition is obtained by just raising the eigenvalues to the same power. The same can be shown for negative powers, so if we want to invert a matrix we need only consider

$$\mathbf{A}^{-1} = \mathbf{W}\boldsymbol{\Sigma}^{-1}\mathbf{W}^{-1}, \tag{A.11}$$

or in other words, just invert each eigenvalue. This will work as long as each eigenvalue is non-zero, so we see that invertible is the same as having no zero eigenvalues.

Indeed, additional work can show that if $\lambda_1, \ldots, \lambda_n$ are the eigenvalues of a matrix, then the determinant of that matrix is

$$\det(\mathbf{A}) = \lambda_1 \cdots \lambda_n, \tag{A.12}$$

or the product of all the eigenvalues. This makes sense intuitively because whatever stretching $\mathbf{W}$ does, $W^{-1}$ undoes it, so in the end the only stretching that happens is by multiplication by the diagonal matrix $\mathbf{\Sigma}$, which stretches volumes by the product of the diagonal elements.

Finally, recall that the rank was the maximum number of linearly independent columns of your matrix. By examining the eigendecomposition closely, we can see that the rank is the same as the number of non-zero eigenvalues of $\mathbf{A}$.

The examples could continue, but hopefully the point is clear: eigendecomposition can simplify many linear-algebraic computations and is a fundamental operation underlying many numerical algorithms and much of the analysis that we do in linear algebra.

### A.2.4 Eigendecompositions of Symmetric Matrices

It is not always possible to find enough linearly independent eigenvectors for the above process to work. For instance the matrix

$$\mathbf{A} = \begin{bmatrix} 1 & 1 \\ 0 & 1 \end{bmatrix}, \tag{A.13}$$

has only a single eigenvector, namely $(1, 0)^\top$. To handle such matrices, we require more advanced techniques than we can cover (such as the Jordan Normal Form, or Singular Value Decomposition). We will often need to restrict our attention to those matrices where we can guarantee the existence of a full set of eigenvectors.

The most commonly encountered family are the *symmetric matrices*, which are those matrices where $\mathbf{A} = \mathbf{A}^\top$. In this case, we may take $W$ to be an *orthogonal matrix*—a matrix whose columns are all length one vectors that are at right angles to one another, where $\mathbf{W}^\top = \mathbf{W}^{-1}$—and all the eigenvalues will be real. Thus, in this special case, we can write (A.9) as

$$\mathbf{A} = \mathbf{W}\mathbf{\Sigma}\mathbf{W}^\top. \tag{A.14}$$

### A.2.5 Gershgorin Circle Theorem

Eigenvalues are often difficult to reason with intuitively. If presented an arbitrary matrix, there is little that can be said about what the eigenvalues are without computing them. There is, however, one theorem that can make it easy to approximate well if the largest values are on the diagonal.

Let $\mathbf{A} = (a_{ij})$ be any square matrix ($n \times n$). We will define $r_i = \sum_{j \neq i} |a_{ij}|$. Let $\mathcal{D}_i$ represent the disc in the complex plane with center $a_{ii}$ radius $r_i$. Then, every eigenvalue of $\mathbf{A}$ is contained in one of the $\mathcal{D}_i$.

This can be a bit to unpack, so let's look at an example. Consider the matrix:

$$\mathbf{A} = \begin{bmatrix} 1.0 & 0.1 & 0.1 & 0.1 \\ 0.1 & 3.0 & 0.2 & 0.3 \\ 0.1 & 0.2 & 5.0 & 0.5 \\ 0.1 & 0.3 & 0.5 & 9.0 \end{bmatrix}. \tag{A.15}$$

We have $r_1 = 0.3$, $r_2 = 0.6$, $r_3 = 0.8$ and $r_4 = 0.9$. The matrix is symmetric, so all eigenvalues are real. This means that all of our eigenvalues will be in one of the ranges of

$$[a_{11} - r_1, a_{11} + r_1] = [0.7, 1.3], \tag{A.16}$$

$$[a_{22} - r_2, a_{22} + r_2] = [2.4, 3.6], \tag{A.17}$$

$$[a_{33} - r_3, a_{33} + r_3] = [4.2, 5.8], \tag{A.18}$$

$$[a_{44} - r_4, a_{44} + r_4] = [8.1, 9.9]. \tag{A.19}$$

Performing the numerical computation shows that the eigenvalues are approximately $0.99$, $2.97$, $4.95$, $9.08$, all comfortably inside the ranges provided.

```
A = torch.tensor([[1.0, 0.1, 0.1, 0.1],
              [0.1, 3.0, 0.2, 0.3],
              [0.1, 0.2, 5.0, 0.5],
              [0.1, 0.3, 0.5, 9.0]])

v, _ = torch.linalg.eig(A)
v
```

```
tensor([0.9923+0.j, 9.0803+0.j, 4.9539+0.j, 2.9734+0.j])
```

In this way, eigenvalues can be approximated, and the approximations will be fairly accurate in the case that the diagonal is significantly larger than all the other elements.

It is a small thing, but with a complex and subtle topic like eigendecomposition, it is good to get any intuitive grasp we can.

## A.2.6 A Useful Application: The Growth of Iterated Maps

Now that we understand what eigenvectors are in principle, let's see how they can be used to provide a deep understanding of a problem central to neural network behavior: proper weight initialization.

### Eigenvectors as Long Term Behavior

The full mathematical investigation of the initialization of deep neural networks is beyond the scope of the text, but we can see a toy version here to understand how eigenvalues can help us see how these models work. As we know, neural networks operate by interspersing layers of linear transformations with non-linear operations. For simplicity here, we will assume that there is no non-linearity, and that the transformation is a single repeated matrix operation $A$, so that the output of our model is

$$\mathbf{v}_{out} = \mathbf{A}\cdot \mathbf{A}\cdots \mathbf{A}\mathbf{v}_{in} = \mathbf{A}^N \mathbf{v}_{in}. \tag{A.20}$$

When these models are initialized, $A$ is taken to be a random matrix with Gaussian entries, so let's make one of those. To be concrete, we start with a mean zero, variance one Gaussian distributed $5 \times 5$ matrix.

```
torch.manual_seed(42)

k = 5
A = torch.randn(k, k, dtype=torch.float64)
A
```

```
tensor([[ 0.2996,  0.2424,  0.2832, -0.2329,  0.6712],
        [ 0.7818, -1.7903, -1.7484,  0.1735, -0.1182],
        [-1.7446, -0.4695,  0.4573,  0.5177, -0.2771],
        [-0.6641,  0.6551,  0.2616, -1.5265, -0.3311],
        [-0.6378,  0.1072,  0.7096,  0.3009, -0.2869]], dtype=torch.float64)
```

## Behavior on Random Data

For simplicity in our toy model, we will assume that the data vector we feed in $\mathbf{v}_{in}$ is a random five dimensional Gaussian vector. Let's think about what we want to have happen. For context, lets think of a generic ML problem, where we are trying to turn input data, like an image, into a prediction, like the probability the image is a picture of a cat. If repeated application of $\mathbf{A}$ stretches a random vector out to be very long, then small changes in input will be amplified into large changes in output—tiny modifications of the input image would lead to vastly different predictions. This does not seem right!

On the flip side, if $\mathbf{A}$ shrinks random vectors to be shorter, then after running through many layers, the vector will essentially shrink to nothing, and the output will not depend on the input. This is also clearly not right either!

We need to walk the narrow line between growth and decay to make sure that our output changes depending on our input, but not much!

Let's see what happens when we repeatedly multiply our matrix $\mathbf{A}$ against a random input vector, and keep track of the norm.

```
# Calculate the sequence of norms after repeatedly applying `A`
v_in = torch.randn(k, 1, dtype=torch.float64)

norm_list = [torch.norm(v_in).item()]
```

```
for i in range(1, 100):
    v_in = A @ v_in
    norm_list.append(torch.norm(v_in).item())

d2l.plot(torch.arange(0, 100), norm_list, 'Iteration', 'Value')
```

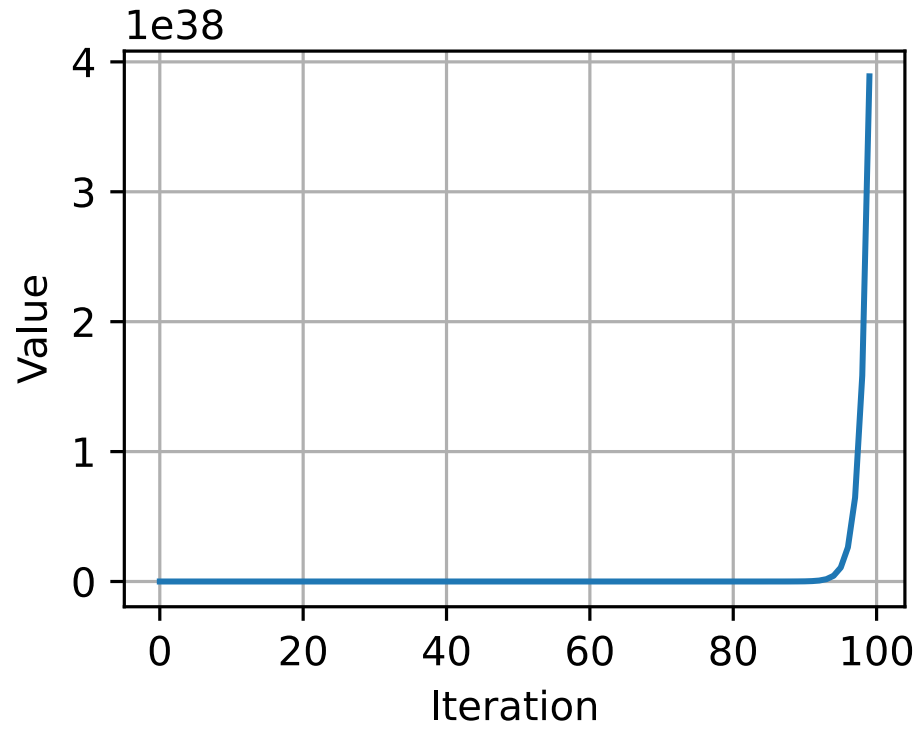

The norm is growing uncontrollably! Indeed if we take the list of quotients, we will see a pattern.

```
# Compute the scaling factor of the norms
norm_ratio_list = []
for i in range(1, 100):
    norm_ratio_list.append(norm_list[i]/norm_list[i - 1])

d2l.plot(torch.arange(1, 100), norm_ratio_list, 'Iteration', 'Ratio')
```

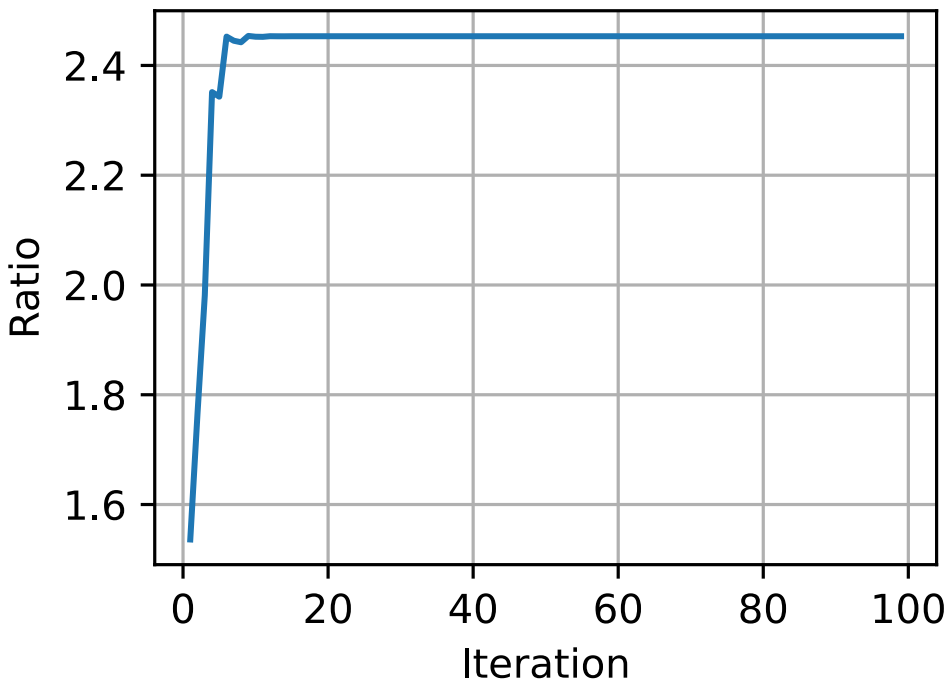

If we look at the last portion of the above computation, we see that the random vector is stretched by a factor of `1.974459321485[...]`, where the portion at the end shifts a little, but the stretching factor is stable.

### Relating Back to Eigenvectors

We have seen that eigenvectors and eigenvalues correspond to the amount something is stretched, but that was for specific vectors, and specific stretches. Let's take a look at what they are for $\mathbf{A}$. A bit of a caveat here: it turns out that to see them all, we will need to go to complex numbers. You can think of these as stretches and rotations. By taking the norm of the complex number (square root of the sums of squares of real and imaginary parts) we can measure that stretching factor. Let's also sort them.

```
# Compute the eigenvalues
eigs = torch.linalg.eig(A).eigenvalues.tolist()
norm_eigs = [torch.abs(torch.tensor(x)) for x in eigs]
norm_eigs.sort()
print(f'norms of eigenvalues: {norm_eigs}')
```

```
norms of eigenvalues: [tensor(0.3490), tensor(1.1296), tensor(1.1296),␣
↪tensor(1.1828), tensor(2.4532)]
```

### An Observation

We see something a bit unexpected happening here: that number we identified before for the long term stretching of our matrix $\mathbf{A}$ applied to a random vector is *exactly* (accurate to thirteen decimal places!) the largest eigenvalue of $\mathbf{A}$. This is clearly not a coincidence!

But, if we now think about what is happening geometrically, this starts to make sense. Consider a random vector. This random vector points a little in every direction, so in particular, it points at least a little bit in the same direction as the eigenvector of $\mathbf{A}$ associated with the largest eigenvalue. This is so important that it is called the *principle eigenvalue* and *principle eigenvector*. After applying $\mathbf{A}$, our random vector gets stretched in every possible direction, as is associated with every possible eigenvector, but it is stretched most of all in the direction associated with this principle eigenvector. What this means is that after apply in $A$, our random vector is longer, and points in a direction closer to being aligned with the principle eigenvector. After applying the matrix many times, the alignment with the principle eigenvector becomes closer and closer until, for all practical purposes, our random vector has been transformed into the principle eigenvector! Indeed this algorithm is the basis for what is known as the *power iteration* for finding the largest eigenvalue and eigenvector of a matrix. For details see, for example, (Golub and Van Loan, 1996).

### Fixing the Normalization

Now, from above discussions, we concluded that we do not want a random vector to be stretched or squished at all, we would like random vectors to stay about the same size throughout the entire process. To do so, we now rescale our matrix by this principle eigenvalue so that the largest eigenvalue is instead now just one. Let's see what happens in this case.

```
# Rescale the matrix `A`
A /= norm_eigs[-1]

# Do the same experiment again
v_in = torch.randn(k, 1, dtype=torch.float64)

norm_list = [torch.norm(v_in).item()]
for i in range(1, 100):
    v_in = A @ v_in
    norm_list.append(torch.norm(v_in).item())

d2l.plot(torch.arange(0, 100), norm_list, 'Iteration', 'Value')
```

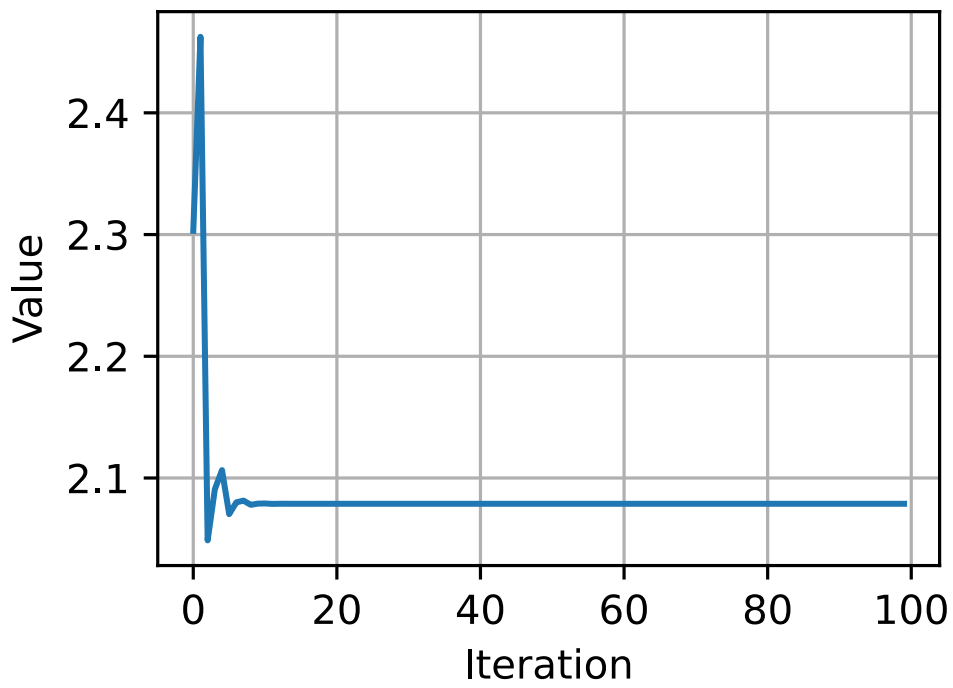

We can also plot the ratio between consecutive norms as before and see that indeed it stabilizes.

```
# Also plot the ratio
norm_ratio_list = []
for i in range(1, 100):
    norm_ratio_list.append(norm_list[i]/norm_list[i-1])

d2l.plot(torch.arange(1, 100), norm_ratio_list, 'Iteration', 'Ratio')
```

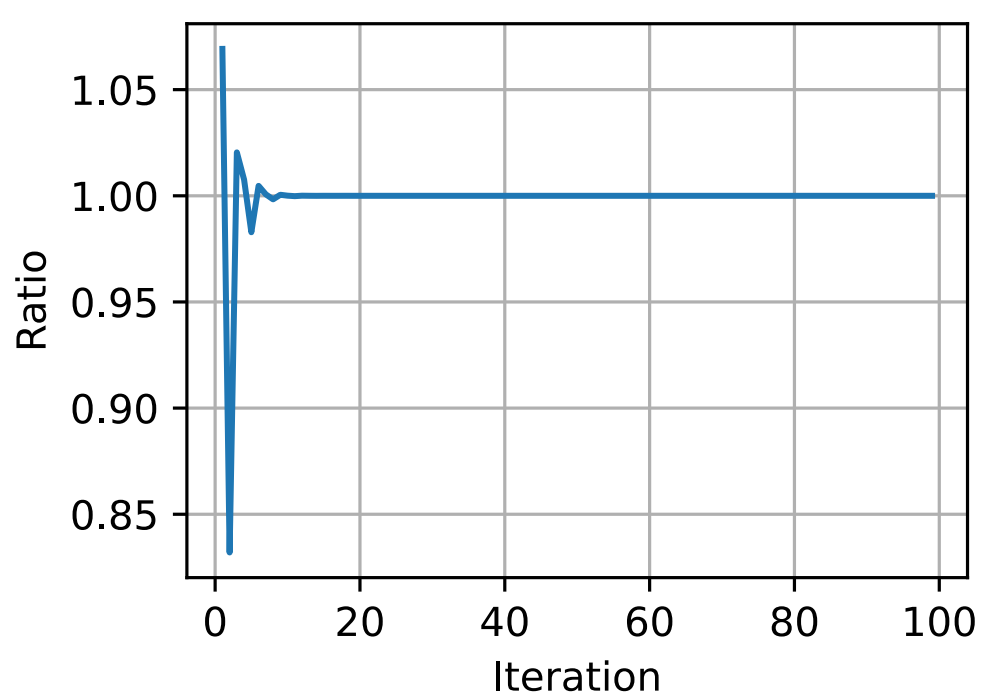

### A.2.7 Discussion

We now see exactly what we hoped for! After normalizing the matrices by the principal eigenvalue, we see that the random data does not explode as before, but rather eventually equilibrates to a specific value. It would be nice to be able to do these things from first principles, and it turns out that if we look deeply at the mathematics of it, we can see that the largest eigenvalue of a large random matrix with independent mean zero, variance one Gaussian entries is on average about $\sqrt{n}$, or in our case $\sqrt{5} \approx 2.2$, due to a fascinating fact known as the *circular law* (Ginibre, 1965). The relationship between the eigenvalues (and a related object called singular values) of random matrices has been shown to have deep connections to proper initialization of neural networks as was discussed in Pennington *et al.* (2017) and subsequent works.

### A.2.8 Summary

- Eigenvectors are vectors which are stretched by a matrix without changing direction.
- Eigenvalues are the amount that the eigenvectors are stretched by the application of the matrix.
- The eigendecomposition of a matrix can allow for many operations to be reduced to operations on the eigenvalues.
- The Gershgorin Circle Theorem can provide approximate values for the eigenvalues of a matrix.
- The behavior of iterated matrix powers depends primarily on the size of the largest eigenvalue. This understanding has many applications in the theory of neural network initialization.

### A.2.9 Exercises

1. What are the eigenvalues and eigenvectors of

$$\mathbf{A} = \begin{bmatrix} 2 & 1 \\ 1 & 2 \end{bmatrix}? \tag{A.21}$$

2. What are the eigenvalues and eigenvectors of the following matrix, and what is strange about this example compared to the previous one?

$$\mathbf{A} = \begin{bmatrix} 2 & 1 \\ 0 & 2 \end{bmatrix}. \tag{A.22}$$

3. Without computing the eigenvalues, is it possible that the smallest eigenvalue of the fol-

lowing matrix is less that $0.5$? *Note*: this problem can be done in your head.

$$\mathbf{A} = \begin{bmatrix} 3.0 & 0.1 & 0.3 & 1.0 \\ 0.1 & 1.0 & 0.1 & 0.2 \\ 0.3 & 0.1 & 5.0 & 0.0 \\ 1.0 & 0.2 & 0.0 & 1.8 \end{bmatrix}. \tag{A.23}$$

Discussions[279].

# A.3 Single Variable Calculus

In Section 2.4, we saw the basic elements of differential calculus. This section takes a deeper dive into the fundamentals of calculus and how we can understand and apply it in the context of machine learning.

## A.3.1 Differential Calculus

Differential calculus is fundamentally the study of how functions behave under small changes. To see why this is so core to deep learning, let's consider an example.

Suppose that we have a deep neural network where the weights are, for convenience, concatenated into a single vector $\mathbf{w} = (w_1, \ldots, w_n)$. Given a training dataset, we consider the loss of our neural network on this dataset, which we will write as $\mathcal{L}(\mathbf{w})$.

This function is extraordinarily complex, encoding the performance of all possible models of the given architecture on this dataset, so it is nearly impossible to tell what set of weights $\mathbf{w}$ will minimize the loss. Thus, in practice, we often start by initializing our weights *randomly*, and then iteratively take small steps in the direction which makes the loss decrease as rapidly as possible.

The question then becomes something that on the surface is no easier: how do we find the direction which makes the weights decrease as quickly as possible? To dig into this, let's first examine the case with only a single weight: $L(\mathbf{w}) = L(x)$ for a single real value $x$.

Let's take $x$ and try to understand what happens when we change it by a small amount to $x + \epsilon$. If you wish to be concrete, think a number like $\epsilon = 0.0000001$. To help us visualize what happens, let's graph an example function, $f(x) = \sin(x^x)$, over the $[0, 3]$.

```
%matplotlib inline
import torch
from IPython import display
from d2l import torch as d2l
```

```
torch.pi = torch.acos(torch.zeros(1)).item() * 2  # Define pi in torch

# Plot a function in a normal range
x_big = torch.arange(0.01, 3.01, 0.01)
ys = torch.sin(x_big**x_big)
d2l.plot(x_big, ys, 'x', 'f(x)')
```

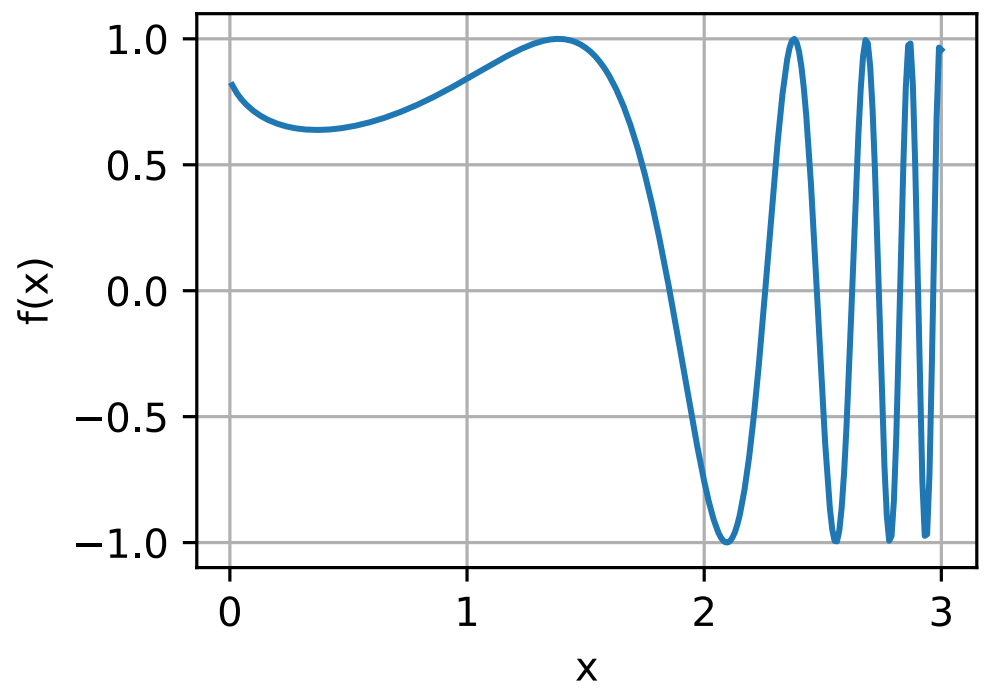

At this large scale, the function's behavior is not simple. However, if we reduce our range to something smaller like $[1.75, 2.25]$, we see that the graph becomes much simpler.

```
# Plot a the same function in a tiny range
x_med = torch.arange(1.75, 2.25, 0.001)
ys = torch.sin(x_med**x_med)
d2l.plot(x_med, ys, 'x', 'f(x)')
```

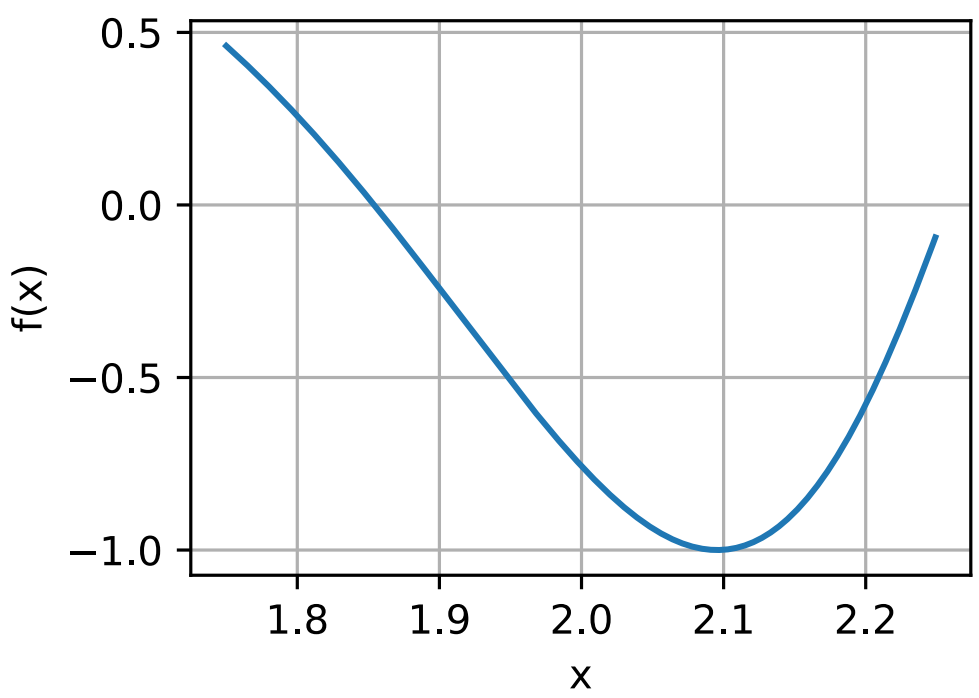

Taking this to an extreme, if we zoom into a tiny segment, the behavior becomes far simpler: it is just a straight line.

```
# Plot a the same function in a tiny range
x_small = torch.arange(2.0, 2.01, 0.0001)
ys = torch.sin(x_small**x_small)
d2l.plot(x_small, ys, 'x', 'f(x)')
```

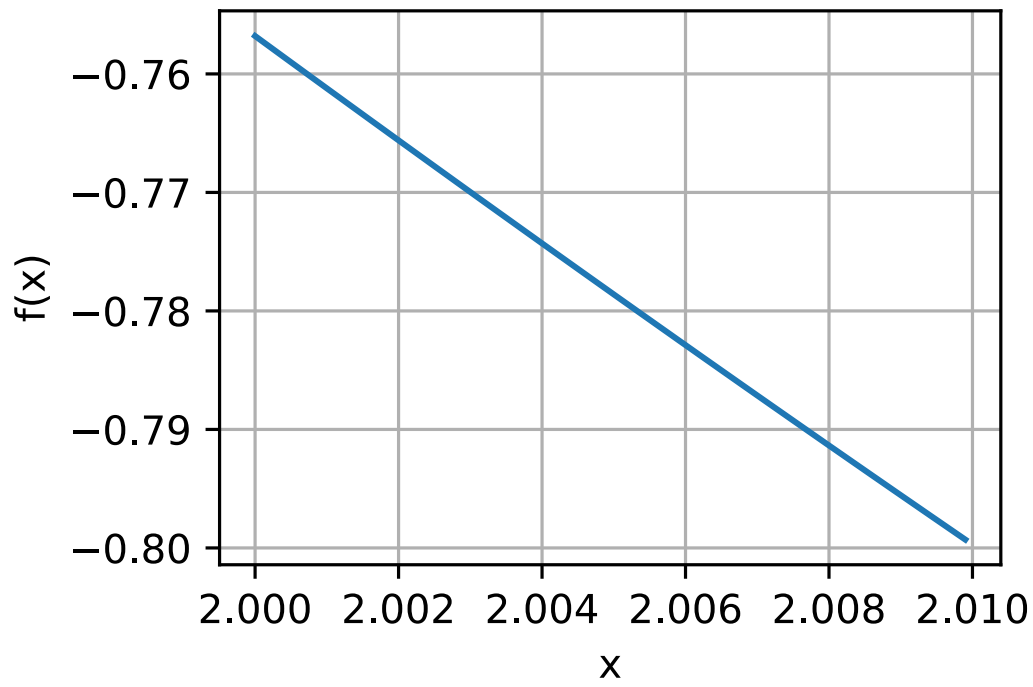

This is the key observation of single variable calculus: the behavior of familiar functions can be modeled by a line in a small enough range. This means that for most functions, it is reasonable to expect that as we shift the $x$ value of the function by a little bit, the output $f(x)$ will also be shifted by a little bit. The only question we need to answer is, "How large is the change in the output compared to the change in the input? Is it half as large? Twice as large?"

Thus, we can consider the ratio of the change in the output of a function for a small change in the input of the function. We can write this formally as

$$\frac{L(x+\epsilon) - L(x)}{(x+\epsilon) - x} = \frac{L(x+\epsilon) - L(x)}{\epsilon}. \tag{A.1}$$

This is already enough to start to play around with in code. For instance, suppose that we know that $L(x) = x^2 + 1701(x-4)^3$, then we can see how large this value is at the point $x = 4$ as follows.

```
# Define our function
def L(x):
    return x**2 + 1701*(x-4)**3

# Print the difference divided by epsilon for several epsilon
for epsilon in [0.1, 0.001, 0.0001, 0.00001]:
    print(f'epsilon = {epsilon:.5f} -> {(L(4+epsilon) - L(4)) / epsilon:.5f}')
```

```
epsilon = 0.10000 -> 25.11000
epsilon = 0.00100 -> 8.00270
epsilon = 0.00010 -> 8.00012
epsilon = 0.00001 -> 8.00001
```

Now, if we are observant, we will notice that the output of this number is suspiciously close to 8. Indeed, if we decrease $\epsilon$, we will see value becomes progressively closer to 8. Thus we may conclude, correctly, that the value we seek (the degree a change in the input changes the output) should be 8 at the point $x = 4$. The way that a mathematician encodes this fact

is

$$\lim_{\epsilon \rightarrow 0} \frac{L(4+\epsilon) - L(4)}{\epsilon} = 8. \tag{A.2}$$

As a bit of a historical digression: in the first few decades of neural network research, scientists used this algorithm (the *method of finite differences*) to evaluate how a loss function changed under small perturbation: just change the weights and see how the loss changed. This is computationally inefficient, requiring two evaluations of the loss function to see how a single change of one variable influenced the loss. If we tried to do this with even a paltry few thousand parameters, it would require several thousand evaluations of the network over the entire dataset! It was not solved until 1986 that the *backpropagation algorithm* introduced in Rumelhart *et al.* (1988) provided a way to calculate how *any* change of the weights together would change the loss in the same computation time as a single prediction of the network over the dataset.

Back in our example, this value 8 is different for different values of $x$, so it makes sense to define it as a function of $x$. More formally, this value dependent rate of change is referred to as the *derivative* which is written as

$$\frac{df}{dx}(x) = \lim_{\epsilon \rightarrow 0} \frac{f(x+\epsilon) - f(x)}{\epsilon}. \tag{A.3}$$

Different texts will use different notations for the derivative. For instance, all of the below notations indicate the same thing:

$$\frac{df}{dx} = \frac{d}{dx}f = f' = \nabla_x f = D_x f = f_x. \tag{A.4}$$

Most authors will pick a single notation and stick with it, however even that is not guaranteed. It is best to be familiar with all of these. We will use the notation $\frac{df}{dx}$ throughout this text, unless we want to take the derivative of a complex expression, in which case we will use $\frac{d}{dx}f$ to write expressions like

$$\frac{d}{dx}\left[x^4 + \cos\left(\frac{x^2+1}{2x-1}\right)\right]. \tag{A.5}$$

Oftentimes, it is intuitively useful to unravel the definition of derivative (A.3) again to see how a function changes when we make a small change of $x$:

$$\begin{aligned}
\frac{df}{dx}(x) = \lim_{\epsilon \rightarrow 0} \frac{f(x+\epsilon) - f(x)}{\epsilon} & \implies \frac{df}{dx}(x) \approx \frac{f(x+\epsilon) - f(x)}{\epsilon} \\
& \implies \epsilon \frac{df}{dx}(x) \approx f(x+\epsilon) - f(x) \\
& \implies f(x+\epsilon) \approx f(x) + \epsilon \frac{df}{dx}(x).
\end{aligned} \tag{A.6}$$

The last equation is worth explicitly calling out. It tells us that if you take any function and change the input by a small amount, the output would change by that small amount scaled by the derivative.

In this way, we can understand the derivative as the scaling factor that tells us how large of change we get in the output from a change in the input.

## A.3.2 Rules of Calculus

We now turn to the task of understanding how to compute the derivative of an explicit function. A full formal treatment of calculus would derive everything from first principles. We will not indulge in this temptation here, but rather provide an understanding of the common rules encountered.

### Common Derivatives

As was seen in Section 2.4, when computing derivatives one can oftentimes use a series of rules to reduce the computation to a few core functions. We repeat them here for ease of reference.

- **Derivative of constants.** $\frac{d}{dx}c = 0$.
- **Derivative of linear functions.** $\frac{d}{dx}(ax) = a$.
- **Power rule.** $\frac{d}{dx}x^n = nx^{n-1}$.
- **Derivative of exponentials.** $\frac{d}{dx}e^x = e^x$.
- **Derivative of the logarithm.** $\frac{d}{dx}\log(x) = \frac{1}{x}$.

### Derivative Rules

If every derivative needed to be separately computed and stored in a table, differential calculus would be near impossible. It is a gift of mathematics that we can generalize the above derivatives and compute more complex derivatives like finding the derivative of $f(x) = \log\left(1 + (x-1)^{10}\right)$. As was mentioned in Section 2.4, the key to doing so is to codify what happens when we take functions and combine them in various ways, most importantly: sums, products, and compositions.

- **Sum rule.** $\frac{d}{dx}\left(g(x) + h(x)\right) = \frac{dg}{dx}(x) + \frac{dh}{dx}(x)$.
- **Product rule.** $\frac{d}{dx}\left(g(x) \cdot h(x)\right) = g(x)\frac{dh}{dx}(x) + \frac{dg}{dx}(x)h(x)$.
- **Chain rule.** $\frac{d}{dx}g(h(x)) = \frac{dg}{dh}(h(x)) \cdot \frac{dh}{dx}(x)$.

Let's see how we may use (A.6) to understand these rules. For the sum rule, consider following chain of reasoning:

$$
\begin{aligned}
f(x+\epsilon) &= g(x+\epsilon) + h(x+\epsilon) \\
&\approx g(x) + \epsilon\frac{dg}{dx}(x) + h(x) + \epsilon\frac{dh}{dx}(x) \\
&= g(x) + h(x) + \epsilon\left(\frac{dg}{dx}(x) + \frac{dh}{dx}(x)\right) \\
&= f(x) + \epsilon\left(\frac{dg}{dx}(x) + \frac{dh}{dx}(x)\right).
\end{aligned} \tag{A.7}
$$

By comparing this result with the fact that $f(x+\epsilon) \approx f(x) + \epsilon \frac{df}{dx}(x)$, we see that $\frac{df}{dx}(x) = \frac{dg}{dx}(x) + \frac{dh}{dx}(x)$ as desired. The intuition here is: when we change the input $x$, $g$ and $h$ jointly contribute to the change of the output by $\frac{dg}{dx}(x)$ and $\frac{dh}{dx}(x)$.

The product is more subtle, and will require a new observation about how to work with these expressions. We will begin as before using (A.6):

$$
\begin{aligned}
f(x+\epsilon) &= g(x+\epsilon) \cdot h(x+\epsilon) \\
&\approx \left(g(x) + \epsilon \frac{dg}{dx}(x)\right) \cdot \left(h(x) + \epsilon \frac{dh}{dx}(x)\right) \\
&= g(x) \cdot h(x) + \epsilon \left(g(x)\frac{dh}{dx}(x) + \frac{dg}{dx}(x)h(x)\right) + \epsilon^2 \frac{dg}{dx}(x)\frac{dh}{dx}(x) \\
&= f(x) + \epsilon \left(g(x)\frac{dh}{dx}(x) + \frac{dg}{dx}(x)h(x)\right) + \epsilon^2 \frac{dg}{dx}(x)\frac{dh}{dx}(x).
\end{aligned} \tag{A.8}
$$

This resembles the computation done above, and indeed we see our answer ($\frac{df}{dx}(x) = g(x)\frac{dh}{dx}(x) + \frac{dg}{dx}(x)h(x)$) sitting next to $\epsilon$, but there is the issue of that term of size $\epsilon^2$. We will refer to this as a *higher-order term*, since the power of $\epsilon^2$ is higher than the power of $\epsilon^1$. We will see in a later section that we will sometimes want to keep track of these, however for now observe that if $\epsilon = 0.0000001$, then $\epsilon^2 = 0.00000000000001$, which is vastly smaller. As we send $\epsilon \rightarrow 0$, we may safely ignore the higher order terms. As a general convention in this appendix, we will use "≈" to denote that the two terms are equal up to higher order terms. However, if we wish to be more formal we may examine the difference quotient

$$
\frac{f(x+\epsilon) - f(x)}{\epsilon} = g(x)\frac{dh}{dx}(x) + \frac{dg}{dx}(x)h(x) + \epsilon \frac{dg}{dx}(x)\frac{dh}{dx}(x), \tag{A.9}
$$

and see that as we send $\epsilon \rightarrow 0$, the right hand term goes to zero as well.

Finally, with the chain rule, we can again progress as before using (A.6) and see that

$$
\begin{aligned}
f(x+\epsilon) &= g(h(x+\epsilon)) \\
&\approx g\left(h(x) + \epsilon \frac{dh}{dx}(x)\right) \\
&\approx g(h(x)) + \epsilon \frac{dh}{dx}(x)\frac{dg}{dh}(h(x)) \\
&= f(x) + \epsilon \frac{dg}{dh}(h(x))\frac{dh}{dx}(x),
\end{aligned} \tag{A.10}
$$

where in the second line we view the function $g$ as having its input ($h(x)$) shifted by the tiny quantity $\epsilon \frac{dh}{dx}(x)$.

These rule provide us with a flexible set of tools to compute essentially any expression desired.

For instance,

$$\begin{aligned}\frac{d}{dx}\left[\log\left(1+(x-1)^{10}\right)\right] &= \left(1+(x-1)^{10}\right)^{-1}\frac{d}{dx}\left[1+(x-1)^{10}\right]\\ &= \left(1+(x-1)^{10}\right)^{-1}\left(\frac{d}{dx}[1]+\frac{d}{dx}[(x-1)^{10}]\right)\\ &= \left(1+(x-1)^{10}\right)^{-1}\left(0+10(x-1)^{9}\frac{d}{dx}[x-1]\right)\\ &= 10\left(1+(x-1)^{10}\right)^{-1}(x-1)^{9}\\ &= \frac{10(x-1)^{9}}{1+(x-1)^{10}}.\end{aligned} \tag{A.11}$$

Where each line has used the following rules:

1. The chain rule and derivative of logarithm.
2. The sum rule.
3. The derivative of constants, chain rule, and power rule.
4. The sum rule, derivative of linear functions, derivative of constants.

Two things should be clear after doing this example:

1. Any function we can write down using sums, products, constants, powers, exponentials, and logarithms can have its derivate computed mechanically by following these rules.
2. Having a human follow these rules can be tedious and error prone!

Thankfully, these two facts together hint towards a way forward: this is a perfect candidate for mechanization! Indeed backpropagation, which we will revisit later in this section, is exactly that.

### Linear Approximation

When working with derivatives, it is often useful to geometrically interpret the approximation used above. In particular, note that the equation

$$f(x+\epsilon) \approx f(x) + \epsilon \frac{df}{dx}(x), \tag{A.12}$$

approximates the value of $f$ by a line which passes through the point $(x, f(x))$ and has slope $\frac{df}{dx}(x)$. In this way we say that the derivative gives a linear approximation to the function $f$, as illustrated below:

```
# Compute sin
xs = torch.arange(-torch.pi, torch.pi, 0.01)
plots = [torch.sin(xs)]

# Compute some linear approximations. Use d(sin(x))/dx = cos(x)
```

```
for x0 in [-1.5, 0.0, 2.0]:
    plots.append(torch.sin(torch.tensor(x0)) + (xs - x0) *
                 torch.cos(torch.tensor(x0)))

d2l.plot(xs, plots, 'x', 'f(x)', ylim=[-1.5, 1.5])
```

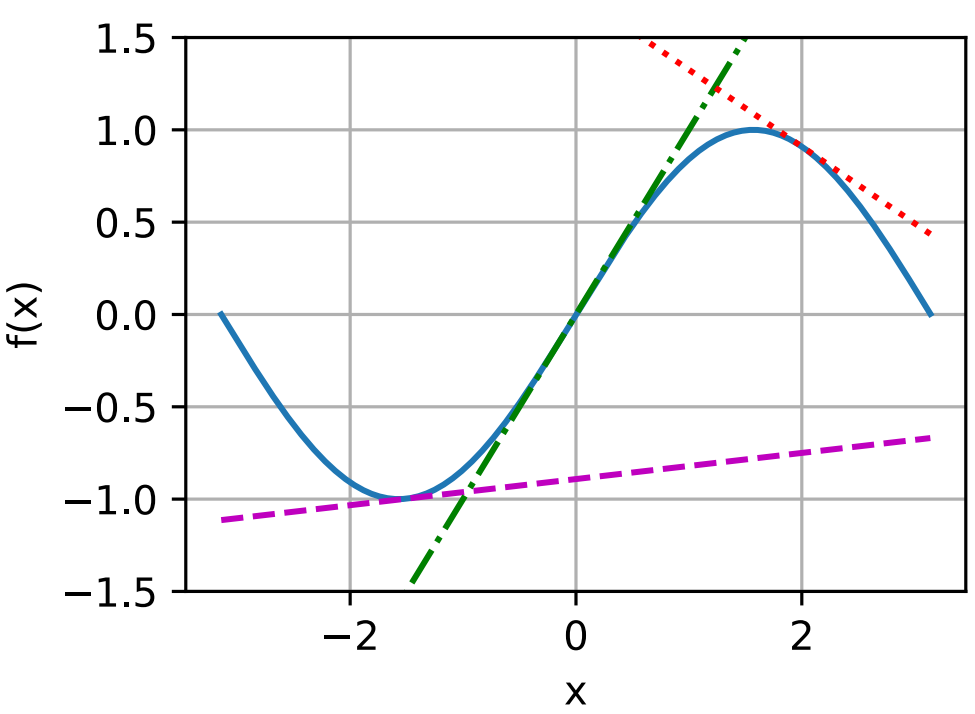

### Higher Order Derivatives

Let's now do something that may on the surface seem strange. Take a function $f$ and compute the derivative $\frac{df}{dx}$. This gives us the rate of change of $f$ at any point.

However, the derivative, $\frac{df}{dx}$, can be viewed as a function itself, so nothing stops us from computing the derivative of $\frac{df}{dx}$ to get $\frac{d^2 f}{dx^2} = \frac{df}{dx}\left(\frac{df}{dx}\right)$. We will call this the second derivative of $f$. This function is the rate of change of the rate of change of $f$, or in other words, how the rate of change is changing. We may apply the derivative any number of times to obtain what is called the $n$-th derivative. To keep the notation clean, we will denote the $n$-th derivative as

$$f^{(n)}(x) = \frac{d^n f}{dx^n} = \left(\frac{d}{dx}\right)^n f. \tag{A.13}$$

Let's try to understand *why* this is a useful notion. Below, we visualize $f^{(2)}(x)$, $f^{(1)}(x)$, and $f(x)$.

First, consider the case that the second derivative $f^{(2)}(x)$ is a positive constant. This means that the slope of the first derivative is positive. As a result, the first derivative $f^{(1)}(x)$ may start out negative, becomes zero at a point, and then becomes positive in the end. This tells us the slope of our original function $f$ and therefore, the function $f$ itself decreases, flattens out, then increases. In other words, the function $f$ curves up, and has a single minimum as is shown in Fig. A.1.

Second, if the second derivative is a negative constant, that means that the first derivative is decreasing. This implies the first derivative may start out positive, becomes zero at a point, and then becomes negative. Hence, the function $f$ itself increases, flattens out, then decreases. In other words, the function $f$ curves down, and has a single maximum as is shown in Fig. A.2.

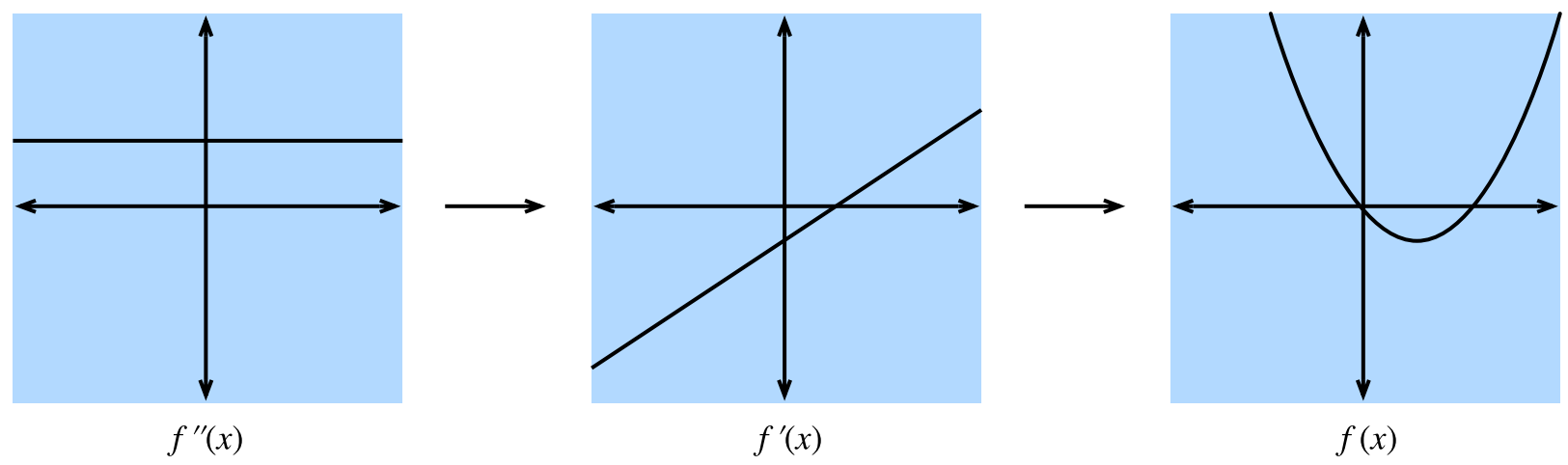

Fig. A.1 If we assume the second derivative is a positive constant, then the fist derivative in increasing, which implies the function itself has a minimum.

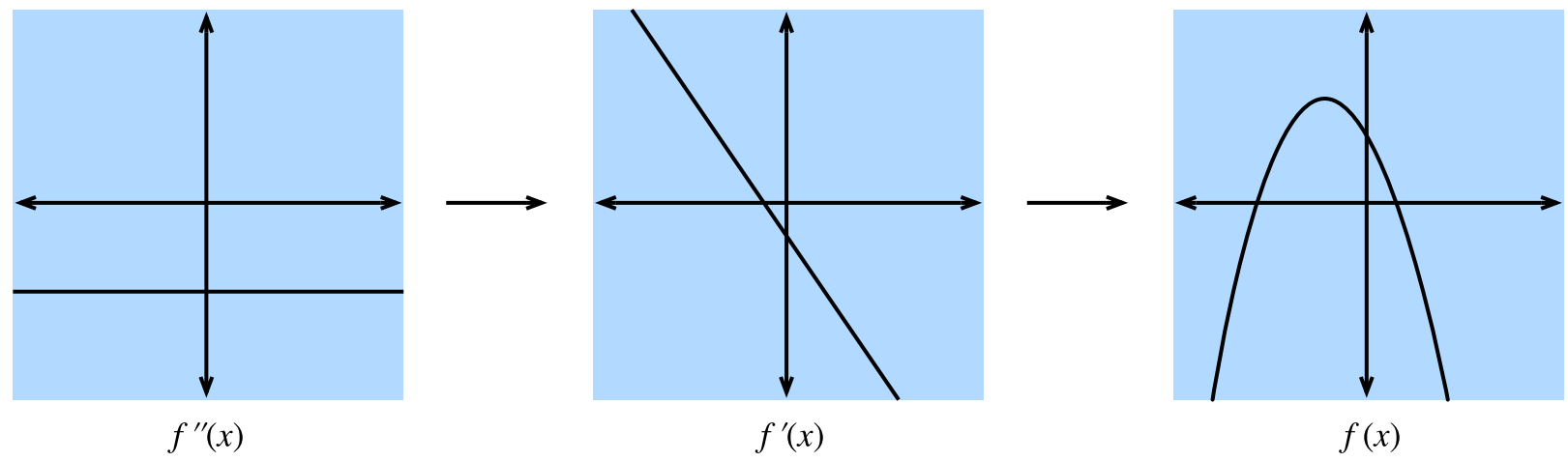

Fig. A.2 If we assume the second derivative is a negative constant, then the fist derivative in decreasing, which implies the function itself has a maximum.

Third, if the second derivative is a always zero, then the first derivative will never change—it is constant! This means that $f$ increases (or decreases) at a fixed rate, and $f$ is itself a straight line as is shown in Fig. A.3.

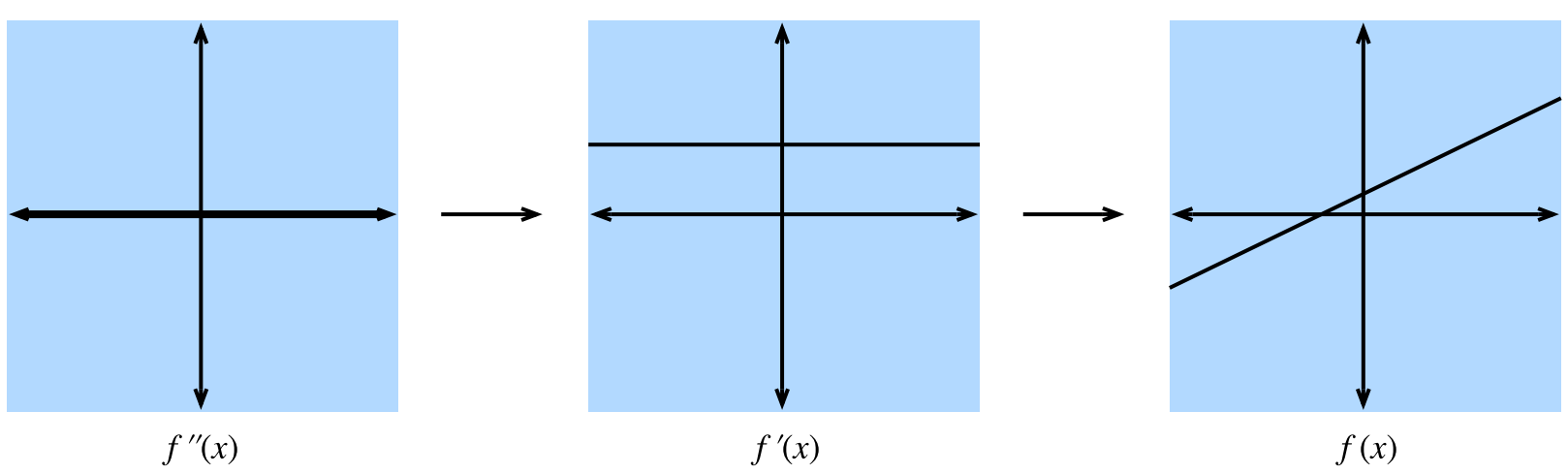

Fig. A.3 If we assume the second derivative is zero, then the fist derivative is constant, which implies the function itself is a straight line.

To summarize, the second derivative can be interpreted as describing the way that the function $f$ curves. A positive second derivative leads to a upwards curve, while a negative second derivative means that $f$ curves downwards, and a zero second derivative means that $f$ does not curve at all.

Let's take this one step further. Consider the function $g(x) = ax^2 + bx + c$. We can then

compute that

$$\begin{aligned}\frac{dg}{dx}(x) &= 2ax + b \\ \frac{d^2g}{dx^2}(x) &= 2a.\end{aligned} \tag{A.14}$$

If we have some original function $f(x)$ in mind, we may compute the first two derivatives and find the values for $a$, $b$, and $c$ that make them match this computation. Similarly to the previous section where we saw that the first derivative gave the best approximation with a straight line, this construction provides the best approximation by a quadratic. Let's visualize this for $f(x) = \sin(x)$.

```
# Compute sin
xs = torch.arange(-torch.pi, torch.pi, 0.01)
plots = [torch.sin(xs)]

# Compute some quadratic approximations. Use d(sin(x)) / dx = cos(x)
for x0 in [-1.5, 0.0, 2.0]:
    plots.append(torch.sin(torch.tensor(x0)) + (xs - x0) *
                 torch.cos(torch.tensor(x0)) - (xs - x0)**2 *
                 torch.sin(torch.tensor(x0)) / 2)

d2l.plot(xs, plots, 'x', 'f(x)', ylim=[-1.5, 1.5])
```

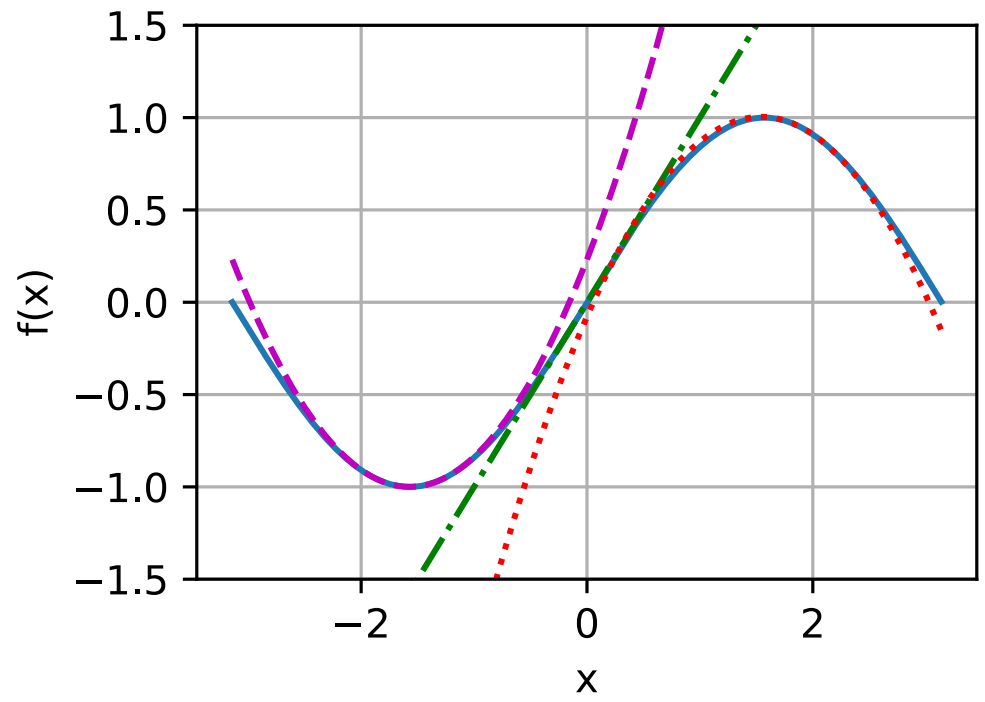

We will extend this idea to the idea of a *Taylor series* in the next section.

## Taylor Series

The *Taylor series* provides a method to approximate the function $f(x)$ if we are given values for the first $n$ derivatives at a point $x_0$, i.e., $\left\{f(x_0), f^{(1)}(x_0), f^{(2)}(x_0), \ldots, f^{(n)}(x_0)\right\}$. The idea will be to find a degree $n$ polynomial that matches all the given derivatives at $x_0$.

We saw the case of $n = 2$ in the previous section and a little algebra shows this is

$$f(x) \approx \frac{1}{2}\frac{d^2f}{dx^2}(x_0)(x-x_0)^2 + \frac{df}{dx}(x_0)(x-x_0) + f(x_0). \tag{A.15}$$

As we can see above, the denominator of 2 is there to cancel out the 2 we get when we take two derivatives of $x^2$, while the other terms are all zero. Same logic applies for the first derivative and the value itself.

If we push the logic further to $n = 3$, we will conclude that

$$f(x) \approx \frac{\frac{d^3f}{dx^3}(x_0)}{6}(x-x_0)^3 + \frac{\frac{d^2f}{dx^2}(x_0)}{2}(x-x_0)^2 + \frac{df}{dx}(x_0)(x-x_0) + f(x_0). \tag{A.16}$$

where the $6 = 3 \times 2 = 3!$ comes from the constant we get in front if we take three derivatives of $x^3$.

Furthermore, we can get a degree $n$ polynomial by

$$P_n(x) = \sum_{i=0}^{n} \frac{f^{(i)}(x_0)}{i!}(x-x_0)^i. \tag{A.17}$$

where the notation

$$f^{(n)}(x) = \frac{d^n f}{dx^n} = \left(\frac{d}{dx}\right)^n f. \tag{A.18}$$

Indeed, $P_n(x)$ can be viewed as the best $n$-th degree polynomial approximation to our function $f(x)$.

While we are not going to dive all the way into the error of the above approximations, it is worth mentioning the infinite limit. In this case, for well behaved functions (known as real analytic functions) like $\cos(x)$ or $e^x$, we can write out the infinite number of terms and approximate the exactly same function

$$f(x) = \sum_{n=0}^{\infty} \frac{f^{(n)}(x_0)}{n!}(x-x_0)^n. \tag{A.19}$$

Take $f(x) = e^x$ as am example. Since $e^x$ is its own derivative, we know that $f^{(n)}(x) = e^x$. Therefore, $e^x$ can be reconstructed by taking the Taylor series at $x_0 = 0$, i.e.,

$$e^x = \sum_{n=0}^{\infty} \frac{x^n}{n!} = 1 + x + \frac{x^2}{2} + \frac{x^3}{6} + \cdots. \tag{A.20}$$

Let's see how this works in code and observe how increasing the degree of the Taylor approximation brings us closer to the desired function $e^x$.

```
# Compute the exponential function
xs = torch.arange(0, 3, 0.01)
ys = torch.exp(xs)

# Compute a few Taylor series approximations
P1 = 1 + xs
P2 = 1 + xs + xs**2 / 2
P5 = 1 + xs + xs**2 / 2 + xs**3 / 6 + xs**4 / 24 + xs**5 / 120

d2l.plot(xs, [ys, P1, P2, P5], 'x', 'f(x)', legend=[
    "Exponential", "Degree 1 Taylor Series", "Degree 2 Taylor Series",
    "Degree 5 Taylor Series"])
```

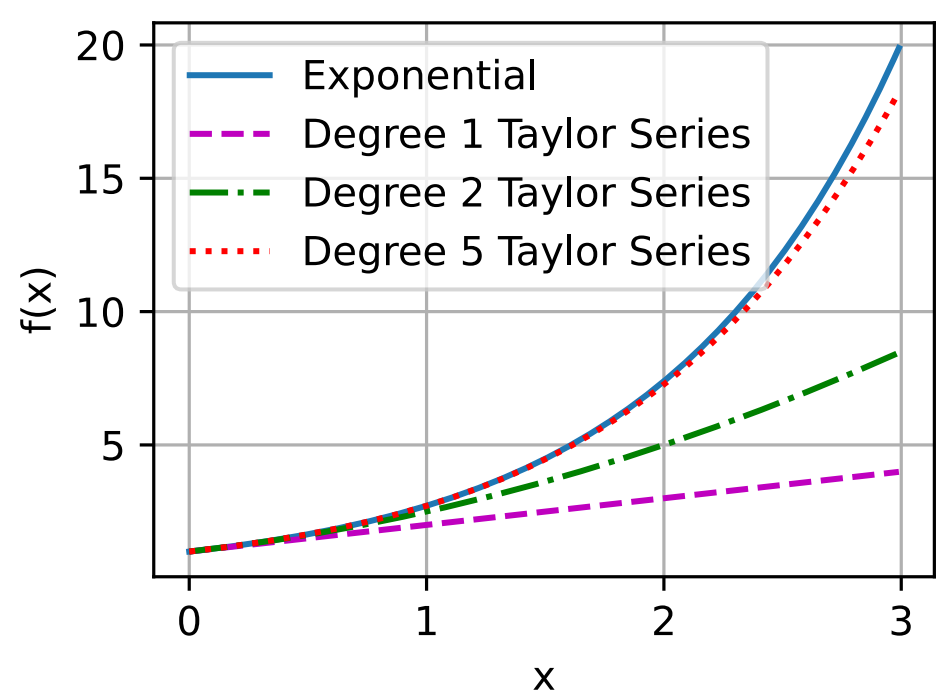

Taylor series have two primary applications:

1. *Theoretical applications*: Often when we try to understand a too complex function, using Taylor series enables us to turn it into a polynomial that we can work with directly.
2. *Numerical applications*: Some functions like $e^x$ or $\cos(x)$ are difficult for machines to compute. They can store tables of values at a fixed precision (and this is often done), but it still leaves open questions like "What is the 1000-th digit of $\cos(1)$?" Taylor series are often helpful to answer such questions.

## A.3.3 Summary

- Derivatives can be used to express how functions change when we change the input by a small amount.
- Elementary derivatives can be combined using derivative rules to create arbitrarily complex derivatives.
- Derivatives can be iterated to get second or higher order derivatives. Each increase in order provides more fine grained information on the behavior of the function.
- Using information in the derivatives of a single data example, we can approximate well behaved functions by polynomials obtained from the Taylor series.

## A.3.4 Exercises

1. What is the derivative of $x^3 - 4x + 1$?
2. What is the derivative of $\log(\frac{1}{x})$?
3. True or False: If $f'(x) = 0$ then $f$ has a maximum or minimum at $x$?
4. Where is the minimum of $f(x) = x\log(x)$ for $x \geq 0$ (where we assume that $f$ takes the limiting value of $0$ at $f(0)$)?

Discussions[280].

# A.4 Multivariable Calculus

Now that we have a fairly strong understanding of derivatives of a function of a single variable, let's return to our original question where we were considering a loss function of potentially billions of weights.

## A.4.1 Higher-Dimensional Differentiation

What Section A.3 tells us is that if we change a single one of these billions of weights leaving every other one fixed, we know what will happen! This is nothing more than a function of a single variable, so we can write

$$L(w_1+\epsilon_1, w_2, \ldots, w_N) \approx L(w_1, w_2, \ldots, w_N) + \epsilon_1 \frac{d}{dw_1} L(w_1, w_2, \ldots, w_N). \tag{A.1}$$

We will call the derivative in one variable while fixing the other variables the *partial derivative*, and we will use the notation $\frac{\partial}{\partial w_1}$ for the derivative in (A.1).

Now, let's take this and change $w_2$ a little bit to $w_2 + \epsilon_2$:

$$\begin{aligned} L(w_1+\epsilon_1, w_2+\epsilon_2, \ldots, w_N) & \approx L(w_1, w_2+\epsilon_2, \ldots, w_N) + \epsilon_1 \frac{\partial}{\partial w_1} L(w_1, w_2+\epsilon_2, \ldots, w_N+\epsilon_N) \\ & \approx L(w_1, w_2, \ldots, w_N) \\ & \quad + \epsilon_2 \frac{\partial}{\partial w_2} L(w_1, w_2, \ldots, w_N) \\ & \quad + \epsilon_1 \frac{\partial}{\partial w_1} L(w_1, w_2, \ldots, w_N) \\ & \quad + \epsilon_1\epsilon_2 \frac{\partial}{\partial w_2}\frac{\partial}{\partial w_1} L(w_1, w_2, \ldots, w_N) \\ & \approx L(w_1, w_2, \ldots, w_N) \\ & \quad + \epsilon_2 \frac{\partial}{\partial w_2} L(w_1, w_2, \ldots, w_N) \\ & \quad + \epsilon_1 \frac{\partial}{\partial w_1} L(w_1, w_2, \ldots, w_N). \end{aligned} \tag{A.2}$$

We have again used the idea that $\epsilon_1\epsilon_2$ is a higher order term that we can discard in the same way we could discard $\epsilon^2$ in the previous section, along with what we saw in (A.1). By continuing in this manner, we may write that

$$L(w_1+\epsilon_1, w_2+\epsilon_2, \ldots, w_N+\epsilon_N) \approx L(w_1, w_2, \ldots, w_N) + \sum_i \epsilon_i \frac{\partial}{\partial w_i} L(w_1, w_2, \ldots, w_N). \tag{A.3}$$

This may look like a mess, but we can make this more familiar by noting that the sum on the right looks exactly like a dot product, so if we let

$$\boldsymbol{\epsilon} = [\epsilon_1, \ldots, \epsilon_N]^\top \text{ and } \nabla_{\mathbf{x}} L = \left[\frac{\partial L}{\partial x_1}, \ldots, \frac{\partial L}{\partial x_N}\right]^\top, \tag{A.4}$$

then

$$L(\mathbf{w} + \boldsymbol{\epsilon}) \approx L(\mathbf{w}) + \boldsymbol{\epsilon} \cdot \nabla_{\mathbf{w}} L(\mathbf{w}). \tag{A.5}$$

We will call the vector $\nabla_{\mathbf{w}} L$ the *gradient* of $L$.

Equation (A.5) is worth pondering for a moment. It has exactly the format that we encountered in one dimension, just we have converted everything to vectors and dot products. It allows us to tell approximately how the function $L$ will change given any perturbation to the input. As we will see in the next section, this will provide us with an important tool in understanding geometrically how we can learn using information contained in the gradient.

But first, let's see this approximation at work with an example. Suppose that we are working with the function

$$f(x, y) = \log(e^x + e^y) \text{ with gradient } \nabla f(x, y) = \left[\frac{e^x}{e^x + e^y}, \frac{e^y}{e^x + e^y}\right]. \tag{A.6}$$

If we look at a point like $(0, \log(2))$, we see that

$$f(x, y) = \log(3) \text{ with gradient } \nabla f(x, y) = \left[\frac{1}{3}, \frac{2}{3}\right]. \tag{A.7}$$

Thus, if we want to approximate $f$ at $(\epsilon_1, \log(2) + \epsilon_2)$, we see that we should have the specific instance of (A.5):

$$f(\epsilon_1, \log(2) + \epsilon_2) \approx \log(3) + \frac{1}{3}\epsilon_1 + \frac{2}{3}\epsilon_2. \tag{A.8}$$

We can test this in code to see how good the approximation is.

```python
%matplotlib inline
import numpy as np
import torch
from IPython import display
from mpl_toolkits import mplot3d
from d2l import torch as d2l

def f(x, y):
    return torch.log(torch.exp(x) + torch.exp(y))
def grad_f(x, y):
    return torch.tensor([torch.exp(x) / (torch.exp(x) + torch.exp(y)),
                         torch.exp(y) / (torch.exp(x) + torch.exp(y))])

epsilon = torch.tensor([0.01, -0.03])
grad_approx = f(torch.tensor([0.]), torch.log(
    torch.tensor([2.]))) + epsilon.dot(
    grad_f(torch.tensor([0.]), torch.log(torch.tensor(2.))))
true_value = f(torch.tensor([0.]) + epsilon[0], torch.log(
    torch.tensor([2.])) + epsilon[1])
f'approximation: {grad_approx}, true Value: {true_value}'
```

```
'approximation: tensor([1.0819]), true Value: tensor([1.0821])'
```

### A.4.2 Geometry of Gradients and Gradient Descent

Consider the expression from (A.5) again:

$$L(\mathbf{w} + \boldsymbol{\epsilon}) \approx L(\mathbf{w}) + \boldsymbol{\epsilon} \cdot \nabla_{\mathbf{w}} L(\mathbf{w}). \tag{A.9}$$

Let's suppose that I want to use this to help minimize our loss $L$. Let's understand geometrically the algorithm of gradient descent first described in Section 2.5. What we will do is the following:

1. Start with a random choice for the initial parameters $\mathbf{w}$.
2. Find the direction $\mathbf{v}$ that makes $L$ decrease the most rapidly at $\mathbf{w}$.
3. Take a small step in that direction: $\mathbf{w} \rightarrow \mathbf{w} + \epsilon\mathbf{v}$.
4. Repeat.

The only thing we do not know exactly how to do is to compute the vector $\mathbf{v}$ in the second step. We will call such a direction the *direction of steepest descent*. Using the geometric understanding of dot products from Section A.1, we see that we can rewrite (A.5) as

$$L(\mathbf{w} + \mathbf{v}) \approx L(\mathbf{w}) + \mathbf{v} \cdot \nabla_{\mathbf{w}} L(\mathbf{w}) = L(\mathbf{w}) + \|\nabla_{\mathbf{w}} L(\mathbf{w})\| \cos(\theta). \tag{A.10}$$

Note that we have taken our direction to have length one for convenience, and used $\theta$ for the angle between $\mathbf{v}$ and $\nabla_{\mathbf{w}} L(\mathbf{w})$. If we want to find the direction that decreases $L$ as rapidly as possible, we want to make this expression as negative as possible. The only way the direction we pick enters into this equation is through $\cos(\theta)$, and thus we wish to make this cosine as negative as possible. Now, recalling the shape of cosine, we can make this as negative as possible by making $\cos(\theta) = -1$ or equivalently making the angle between the gradient and our chosen direction to be $\pi$ radians, or equivalently $180$ degrees. The only way to achieve this is to head in the exact opposite direction: pick $\mathbf{v}$ to point in the exact opposite direction to $\nabla_{\mathbf{w}} L(\mathbf{w})$!

This brings us to one of the most important mathematical concepts in machine learning: the direction of steepest decent points in the direction of $-\nabla_{\mathbf{w}} L(\mathbf{w})$. Thus our informal algorithm can be rewritten as follows.

1. Start with a random choice for the initial parameters $\mathbf{w}$.
2. Compute $\nabla_{\mathbf{w}} L(\mathbf{w})$.
3. Take a small step in the opposite of that direction: $\mathbf{w} \leftarrow \mathbf{w} - \epsilon\nabla_{\mathbf{w}} L(\mathbf{w})$.
4. Repeat.

This basic algorithm has been modified and adapted many ways by many researchers, but the core concept remains the same in all of them. Use the gradient to find the direction that decreases the loss as rapidly as possible, and update the parameters to take a step in that direction.

### A.4.3 A Note on Mathematical Optimization

Throughout this book, we focus squarely on numerical optimization techniques for the practical reason that all functions we encounter in the deep learning setting are too complex to minimize explicitly.

However, it is a useful exercise to consider what the geometric understanding we obtained above tells us about optimizing functions directly.

Suppose that we wish to find the value of $\mathbf{x}_0$ which minimizes some function $L(\mathbf{x})$. Let's suppose that moreover someone gives us a value and tells us that it is the value that minimizes $L$. Is there anything we can check to see if their answer is even plausible?

Again consider (A.5):

$$L(\mathbf{x}_0 + \boldsymbol{\epsilon}) \approx L(\mathbf{x}_0) + \boldsymbol{\epsilon} \cdot \nabla_{\mathbf{x}} L(\mathbf{x}_0). \tag{A.11}$$

If the gradient is not zero, we know that we can take a step in the direction $-\epsilon \nabla_{\mathbf{x}} L(\mathbf{x}_0)$ to find a value of $L$ that is smaller. Thus, if we truly are at a minimum, this cannot be the case! We can conclude that if $\mathbf{x}_0$ is a minimum, then $\nabla_{\mathbf{x}} L(\mathbf{x}_0) = 0$. We call points with $\nabla_{\mathbf{x}} L(\mathbf{x}_0) = 0$ *critical points*.

This is nice, because in some rare settings, we *can* explicitly find all the points where the gradient is zero, and find the one with the smallest value.

For a concrete example, consider the function

$$f(x) = 3x^4 - 4x^3 - 12x^2. \tag{A.12}$$

This function has derivative

$$\frac{df}{dx} = 12x^3 - 12x^2 - 24x = 12x(x-2)(x+1). \tag{A.13}$$

The only possible location of minima are at $x = -1, 0, 2$, where the function takes the values $-5, 0, -32$ respectively, and thus we can conclude that we minimize our function when $x = 2$. A quick plot confirms this.

```
x = torch.arange(-2, 3, 0.01)
f = (3 * x**4) - (4 * x**3) - (12 * x**2)

d2l.plot(x, f, 'x', 'f(x)')
```

This highlights an important fact to know when working either theoretically or numerically: the only possible points where we can minimize (or maximize) a function will have gradient equal to zero, however, not every point with gradient zero is the true *global* minimum (or maximum).

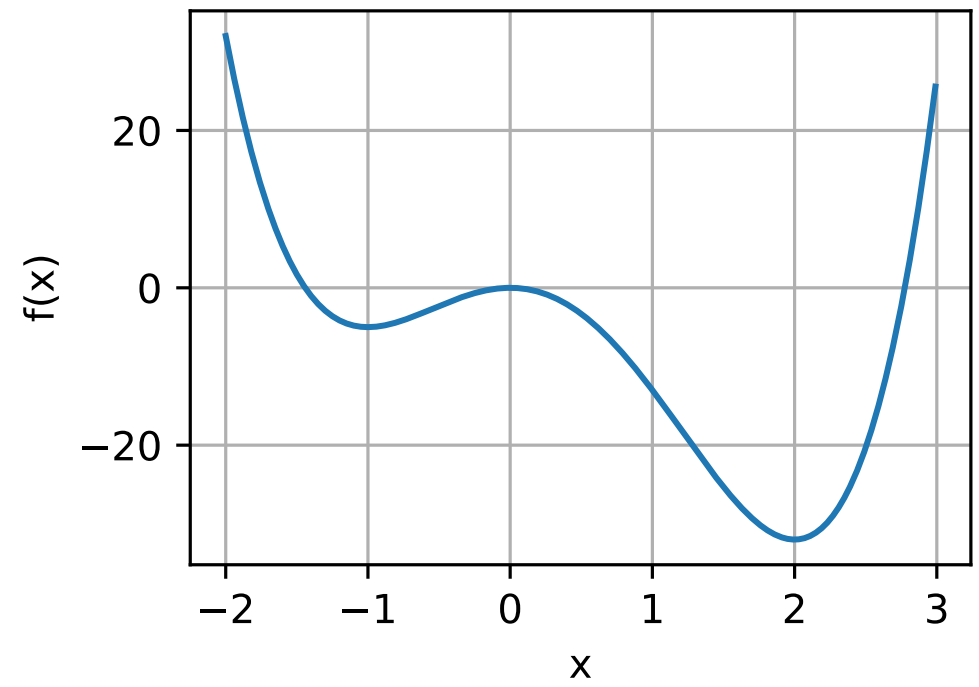

## A.4.4 Multivariate Chain Rule

Let's suppose that we have a function of four variables ($w$, $x$, $y$, and $z$) which we can make by composing many terms:

$$\begin{aligned} f(u, v) &= (u+v)^2 \\ u(a, b) &= (a+b)^2, \qquad v(a, b) = (a-b)^2, \\ a(w, x, y, z) &= (w+x+y+z)^2, \qquad b(w, x, y, z) = (w+x-y-z)^2. \end{aligned} \tag{A.14}$$

Such chains of equations are common when working with neural networks, so trying to understand how to compute gradients of such functions is key. We can start to see visual hints of this connection in Fig. A.1 if we take a look at what variables directly relate to one another.

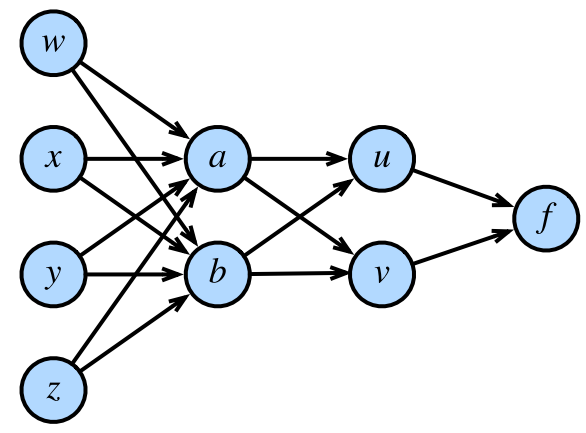

Fig. A.1 The function relations above where nodes represent values and edges show functional dependence.

Nothing stops us from just composing everything from (A.14) and writing out that

$$f(w, x, y, z) = \left(\left((w+x+y+z)^2+(w+x-y-z)^2\right)^2+\left((w+x+y+z)^2-(w+x-y-z)^2\right)^2\right)^2. \tag{A.15}$$

We may then take the derivative by just using single variable derivatives, but if we did that we would quickly find ourself swamped with terms, many of which are repeats! Indeed, one can see that, for instance:

$$\begin{aligned}\frac{\partial f}{\partial w} = & 2\left(2\left(2(w+x+y+z)-2(w+x-y-z)\right)\left((w+x+y+z)^2-(w+x-y-z)^2\right)+\right.\\ & 2\left(2(w+x-y-z)+2(w+x+y+z)\right)\left((w+x-y-z)^2+(w+x+y+z)^2\right)\right)\times\\ & \left(\left((w+x+y+z)^2-(w+x-y-z)^2\right)^2+\left((w+x-y-z)^2+(w+x+y+z)^2\right)^2\right).\end{aligned} \tag{A.16}$$

If we then also wanted to compute $\frac{\partial f}{\partial x}$, we would end up with a similar equation again with many repeated terms, and many *shared* repeated terms between the two derivatives. This represents a massive quantity of wasted work, and if we needed to compute derivatives this way, the whole deep learning revolution would have stalled out before it began!

Let's break up the problem. We will start by trying to understand how $f$ changes when we change $a$, essentially assuming that $w, x, y$, and $z$ all do not exist. We will reason as we did back when we worked with the gradient for the first time. Let's take $a$ and add a small amount $\epsilon$ to it.

$$\begin{aligned} & f(u(a+\epsilon, b), v(a+\epsilon, b)) \\ \approx & f\left(u(a, b)+\epsilon \frac{\partial u}{\partial a}(a, b), v(a, b)+\epsilon \frac{\partial v}{\partial a}(a, b)\right) \\ \approx & f(u(a, b), v(a, b))+\epsilon\left[\frac{\partial f}{\partial u}(u(a, b), v(a, b)) \frac{\partial u}{\partial a}(a, b)+\frac{\partial f}{\partial v}(u(a, b), v(a, b)) \frac{\partial v}{\partial a}(a, b)\right].\end{aligned} \tag{A.17}$$

The first line follows from the definition of partial derivative, and the second follows from the definition of gradient. It is notationally burdensome to track exactly where we evaluate every derivative, as in the expression $\frac{\partial f}{\partial u}(u(a, b), v(a, b))$, so we often abbreviate this to the much more memorable

$$\frac{\partial f}{\partial a}=\frac{\partial f}{\partial u} \frac{\partial u}{\partial a}+\frac{\partial f}{\partial v} \frac{\partial v}{\partial a}. \tag{A.18}$$

It is useful to think about the meaning of the process. We are trying to understand how a function of the form $f(u(a, b), v(a, b))$ changes its value with a change in $a$. There are two pathways this can occur: there is the pathway where $a \rightarrow u \rightarrow f$ and where $a \rightarrow v \rightarrow f$. We can compute both of these contributions via the chain rule: $\frac{\partial w}{\partial u} \cdot \frac{\partial u}{\partial x}$ and $\frac{\partial w}{\partial v} \cdot \frac{\partial v}{\partial x}$ respectively, and added up.

Imagine we have a different network of functions where the functions on the right depend on those that are connected to on the left as is shown in Fig. A.2.

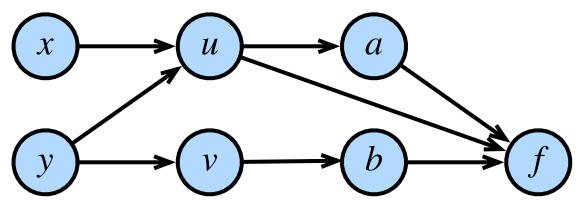

Fig. A.2 Another more subtle example of the chain rule.

To compute something like $\frac{\partial f}{\partial y}$, we need to sum over all (in this case 3) paths from $y$ to $f$

giving

$$\frac{\partial f}{\partial y} = \frac{\partial f}{\partial a}\frac{\partial a}{\partial u}\frac{\partial u}{\partial y} + \frac{\partial f}{\partial u}\frac{\partial u}{\partial y} + \frac{\partial f}{\partial b}\frac{\partial b}{\partial v}\frac{\partial v}{\partial y}. \tag{A.19}$$

Understanding the chain rule in this way will pay great dividends when trying to understand how gradients flow through networks, and why various architectural choices like those in LSTMs (Section 10.1) or residual layers (Section 8.6) can help shape the learning process by controlling gradient flow.

### A.4.5 The Backpropagation Algorithm

Let's return to the example of (A.14) the previous section where

$$\begin{aligned} f(u, v) & = (u+v)^{2} \\ u(a, b) & = (a+b)^{2}, \qquad v(a, b) = (a-b)^{2}, \\ a(w, x, y, z) & = (w+x+y+z)^{2}, \qquad b(w, x, y, z) = (w+x-y-z)^2. \end{aligned} \tag{A.20}$$

If we want to compute say $\frac{\partial f}{\partial w}$ we may apply the multi-variate chain rule to see:

$$\begin{aligned} \frac{\partial f}{\partial w} & = \frac{\partial f}{\partial u}\frac{\partial u}{\partial w} + \frac{\partial f}{\partial v}\frac{\partial v}{\partial w}, \\ \frac{\partial u}{\partial w} & = \frac{\partial u}{\partial a}\frac{\partial a}{\partial w}+\frac{\partial u}{\partial b}\frac{\partial b}{\partial w}, \\ \frac{\partial v}{\partial w} & = \frac{\partial v}{\partial a}\frac{\partial a}{\partial w}+\frac{\partial v}{\partial b}\frac{\partial b}{\partial w}. \end{aligned} \tag{A.21}$$

Let's try using this decomposition to compute $\frac{\partial f}{\partial w}$. Notice that all we need here are the various single step partials:

$$\begin{aligned} \frac{\partial f}{\partial u} = 2(u+v), & \quad \frac{\partial f}{\partial v} = 2(u+v), \\ \frac{\partial u}{\partial a} = 2(a+b), & \quad \frac{\partial u}{\partial b} = 2(a+b), \\ \frac{\partial v}{\partial a} = 2(a-b), & \quad \frac{\partial v}{\partial b} = -2(a-b), \\ \frac{\partial a}{\partial w} = 2(w+x+y+z), & \quad \frac{\partial b}{\partial w} = 2(w+x-y-z). \end{aligned} \tag{A.22}$$

If we write this out into code this becomes a fairly manageable expression.

```
# Compute the value of the function from inputs to outputs
w, x, y, z = -1, 0, -2, 1
a, b = (w + x + y + z)**2, (w + x - y - z)**2
u, v = (a + b)**2, (a - b)**2
f = (u + v)**2
print(f'    f at {w}, {x}, {y}, {z} is {f}')

# Compute the single step partials
df_du, df_dv = 2*(u + v), 2*(u + v)
du_da, du_db, dv_da, dv_db = 2*(a + b), 2*(a + b), 2*(a - b), -2*(a - b)
```

```
da_dw, db_dw = 2*(w + x + y + z), 2*(w + x - y - z)

# Compute the final result from inputs to outputs
du_dw, dv_dw = du_da*da_dw + du_db*db_dw, dv_da*da_dw + dv_db*db_dw
df_dw = df_du*du_dw + df_dv*dv_dw
print(f'df/dw at {w}, {x}, {y}, {z} is {df_dw}')
```

```
    f at -1, 0, -2, 1 is 1024
df/dw at -1, 0, -2, 1 is -4096
```

However, note that this still does not make it easy to compute something like $\frac{\partial f}{\partial x}$. The reason for that is the *way* we chose to apply the chain rule. If we look at what we did above, we always kept $\partial w$ in the denominator when we could. In this way, we chose to apply the chain rule seeing how $w$ changed every other variable. If that is what we wanted, this would be a good idea. However, think back to our motivation from deep learning: we want to see how every parameter changes the *loss*. In essence, we want to apply the chain rule keeping $\partial f$ in the numerator whenever we can!

To be more explicit, note that we can write

$$
\begin{aligned}
\frac{\partial f}{\partial w} & = \frac{\partial f}{\partial a}\frac{\partial a}{\partial w} + \frac{\partial f}{\partial b}\frac{\partial b}{\partial w}, \\
\frac{\partial f}{\partial a} & = \frac{\partial f}{\partial u}\frac{\partial u}{\partial a} + \frac{\partial f}{\partial v}\frac{\partial v}{\partial a}, \\
\frac{\partial f}{\partial b} & = \frac{\partial f}{\partial u}\frac{\partial u}{\partial b} + \frac{\partial f}{\partial v}\frac{\partial v}{\partial b}.
\end{aligned}
\tag{A.23}
$$

Note that this application of the chain rule has us explicitly compute $\frac{\partial f}{\partial u}, \frac{\partial f}{\partial v}, \frac{\partial f}{\partial a}, \frac{\partial f}{\partial b}$, and $\frac{\partial f}{\partial w}$. Nothing stops us from also including the equations:

$$
\begin{aligned}
\frac{\partial f}{\partial x} & = \frac{\partial f}{\partial a}\frac{\partial a}{\partial x} + \frac{\partial f}{\partial b}\frac{\partial b}{\partial x}, \\
\frac{\partial f}{\partial y} & = \frac{\partial f}{\partial a}\frac{\partial a}{\partial y} + \frac{\partial f}{\partial b}\frac{\partial b}{\partial y}, \\
\frac{\partial f}{\partial z} & = \frac{\partial f}{\partial a}\frac{\partial a}{\partial z} + \frac{\partial f}{\partial b}\frac{\partial b}{\partial z}.
\end{aligned}
\tag{A.24}
$$

and then keeping track of how $f$ changes when we change *any* node in the entire network. Let's implement it.

```
# Compute the value of the function from inputs to outputs
w, x, y, z = -1, 0, -2, 1
a, b = (w + x + y + z)**2, (w + x - y - z)**2
u, v = (a + b)**2, (a - b)**2
f = (u + v)**2
print(f'f at {w}, {x}, {y}, {z} is {f}')

# Compute the derivative using the decomposition above
# First compute the single step partials
```

```
df_du, df_dv = 2*(u + v), 2*(u + v)
du_da, du_db, dv_da, dv_db = 2*(a + b), 2*(a + b), 2*(a - b), -2*(a - b)
da_dw, db_dw = 2*(w + x + y + z), 2*(w + x - y - z)
da_dx, db_dx = 2*(w + x + y + z), 2*(w + x - y - z)
da_dy, db_dy = 2*(w + x + y + z), -2*(w + x - y - z)
da_dz, db_dz = 2*(w + x + y + z), -2*(w + x - y - z)

# Now compute how f changes when we change any value from output to input
df_da, df_db = df_du*du_da + df_dv*dv_da, df_du*du_db + df_dv*dv_db
df_dw, df_dx = df_da*da_dw + df_db*db_dw, df_da*da_dx + df_db*db_dx
df_dy, df_dz = df_da*da_dy + df_db*db_dy, df_da*da_dz + df_db*db_dz

print(f'df/dw at {w}, {x}, {y}, {z} is {df_dw}')
print(f'df/dx at {w}, {x}, {y}, {z} is {df_dx}')
print(f'df/dy at {w}, {x}, {y}, {z} is {df_dy}')
print(f'df/dz at {w}, {x}, {y}, {z} is {df_dz}')
```

```
f at -1, 0, -2, 1 is 1024
df/dw at -1, 0, -2, 1 is -4096
df/dx at -1, 0, -2, 1 is -4096
df/dy at -1, 0, -2, 1 is -4096
df/dz at -1, 0, -2, 1 is -4096
```

The fact that we compute derivatives from $f$ back towards the inputs rather than from the inputs forward to the outputs (as we did in the first code snippet above) is what gives this algorithm its name: *backpropagation*. Note that there are two steps: 1. Compute the value of the function, and the single step partials from front to back. While not done above, this can be combined into a single *forward pass*. 2. Compute the gradient of $f$ from back to front. We call this the *backwards pass*.

This is precisely what every deep learning algorithm implements to allow the computation of the gradient of the loss with respect to every weight in the network at one pass. It is an astonishing fact that we have such a decomposition.

To see how to encapsulated this, let's take a quick look at this example.

```
# Initialize as ndarrays, then attach gradients
w = torch.tensor([-1.], requires_grad=True)
x = torch.tensor([0.], requires_grad=True)
y = torch.tensor([-2.], requires_grad=True)
z = torch.tensor([1.], requires_grad=True)
# Do the computation like usual, tracking gradients
a, b = (w + x + y + z)**2, (w + x - y - z)**2
u, v = (a + b)**2, (a - b)**2
f = (u + v)**2

# Execute backward pass
f.backward()

print(f'df/dw at {w.data.item()}, {x.data.item()}, {y.data.item()}, '
      f'{z.data.item()} is {w.grad.data.item()}')
```

```
print(f'df/dx at {w.data.item()}, {x.data.item()}, {y.data.item()}, '
      f'{z.data.item()} is {x.grad.data.item()}')
print(f'df/dy at {w.data.item()}, {x.data.item()}, {y.data.item()}, '
      f'{z.data.item()} is {y.grad.data.item()}')
print(f'df/dz at {w.data.item()}, {x.data.item()}, {y.data.item()}, '
      f'{z.data.item()} is {z.grad.data.item()}')
```

```
df/dw at -1.0, 0.0, -2.0, 1.0 is -4096.0
df/dx at -1.0, 0.0, -2.0, 1.0 is -4096.0
df/dy at -1.0, 0.0, -2.0, 1.0 is -4096.0
df/dz at -1.0, 0.0, -2.0, 1.0 is -4096.0
```

All of what we did above can be done automatically by calling `f.backwards()`.

### A.4.6 Hessians

As with single variable calculus, it is useful to consider higher-order derivatives in order to get a handle on how we can obtain a better approximation to a function than using the gradient alone.

There is one immediate problem one encounters when working with higher order derivatives of functions of several variables, and that is there are a large number of them. If we have a function $f(x_1, \ldots, x_n)$ of $n$ variables, then we can take $n^2$ many second derivatives, namely for any choice of $i$ and $j$:

$$\frac{d^2 f}{dx_i dx_j} = \frac{d}{dx_i}\left(\frac{d}{dx_j} f\right). \tag{A.25}$$

This is traditionally assembled into a matrix called the *Hessian*:

$$\mathbf{H}_f = \begin{bmatrix} \frac{d^2 f}{dx_1 dx_1} & \cdots & \frac{d^2 f}{dx_1 dx_n} \\ \vdots & \ddots & \vdots \\ \frac{d^2 f}{dx_n dx_1} & \cdots & \frac{d^2 f}{dx_n dx_n} \end{bmatrix}. \tag{A.26}$$

Not every entry of this matrix is independent. Indeed, we can show that as long as both *mixed partials* (partial derivatives with respect to more than one variable) exist and are continuous, we can say that for any $i$, and $j$,

$$\frac{d^2 f}{dx_i dx_j} = \frac{d^2 f}{dx_j dx_i}. \tag{A.27}$$

This follows by considering first perturbing a function in the direction of $x_i$, and then perturbing it in $x_j$ and then comparing the result of that with what happens if we perturb first $x_j$ and then $x_i$, with the knowledge that both of these orders lead to the same final change in the output of $f$.

As with single variables, we can use these derivatives to get a far better idea of how the function behaves near a point. In particular, we can use it to find the best fitting quadratic near a point $\mathbf{x}_0$, as we saw in a single variable.

Let's see an example. Suppose that $f(x_1, x_2) = a + b_1 x_1 + b_2 x_2 + c_{11} x_1^2 + c_{12} x_1 x_2 + c_{22} x_2^2$. This is the general form for a quadratic in two variables. If we look at the value of the function, its gradient, and its Hessian (A.26), all at the point zero:

$$
\begin{aligned}
f(0,0) &= a, \\
\nabla f(0,0) &= \begin{bmatrix} b_1 \\ b_2 \end{bmatrix}, \\
\mathbf{H} f(0,0) &= \begin{bmatrix} 2c_{11} & c_{12} \\ c_{12} & 2c_{22} \end{bmatrix},
\end{aligned}
\tag{A.28}
$$

we can get our original polynomial back by saying

$$
f(\mathbf{x}) = f(0) + \nabla f(0) \cdot \mathbf{x} + \frac{1}{2}\mathbf{x}^\top \mathbf{H} f(0) \mathbf{x}. \tag{A.29}
$$

In general, if we computed this expansion any point $\mathbf{x}_0$, we see that

$$
f(\mathbf{x}) = f(\mathbf{x}_0) + \nabla f(\mathbf{x}_0) \cdot (\mathbf{x}-\mathbf{x}_0) + \frac{1}{2}(\mathbf{x}-\mathbf{x}_0)^\top \mathbf{H} f(\mathbf{x}_0)(\mathbf{x}-\mathbf{x}_0). \tag{A.30}
$$

This works for any dimensional input, and provides the best approximating quadratic to any function at a point. To give an example, let's plot the function

$$
f(x, y) = xe^{-x^2-y^2}. \tag{A.31}
$$

One can compute that the gradient and Hessian are

$$
\nabla f(x, y) = e^{-x^2-y^2}\begin{pmatrix}1 - 2x^2 \\ -2xy\end{pmatrix} \text{ and } \mathbf{H}f(x, y) = e^{-x^2-y^2}\begin{pmatrix} 4x^3 - 6x & 4x^2y - 2y \\ 4x^2y-2y & 4xy^2-2x\end{pmatrix}. \tag{A.32}
$$

And thus, with a little algebra, see that the approximating quadratic at $[-1,0]^\top$ is

$$
f(x, y) \approx e^{-1}\left(-1 - (x+1) + (x+1)^2 + y^2\right). \tag{A.33}
$$

```python
# Construct grid and compute function
x, y = torch.meshgrid(torch.linspace(-2, 2, 101),
                   torch.linspace(-2, 2, 101))

z = x*torch.exp(- x**2 - y**2)

# Compute approximating quadratic with gradient and Hessian at (1, 0)
w = torch.exp(torch.tensor([-1.]))*(-1 - (x + 1) + 2 * (x + 1)**2 + 2 * y**2)

# Plot function
ax = d2l.plt.figure().add_subplot(111, projection='3d')
ax.plot_wireframe(x.numpy(), y.numpy(), z.numpy(),
                  **{'rstride': 10, 'cstride': 10})
ax.plot_wireframe(x.numpy(), y.numpy(), w.numpy(),
                  **{'rstride': 10, 'cstride': 10}, color='purple')
```

```
d2l.plt.xlabel('x')
d2l.plt.ylabel('y')
d2l.set_figsize()
ax.set_xlim(-2, 2)
ax.set_ylim(-2, 2)
ax.set_zlim(-1, 1)
ax.dist = 12
```

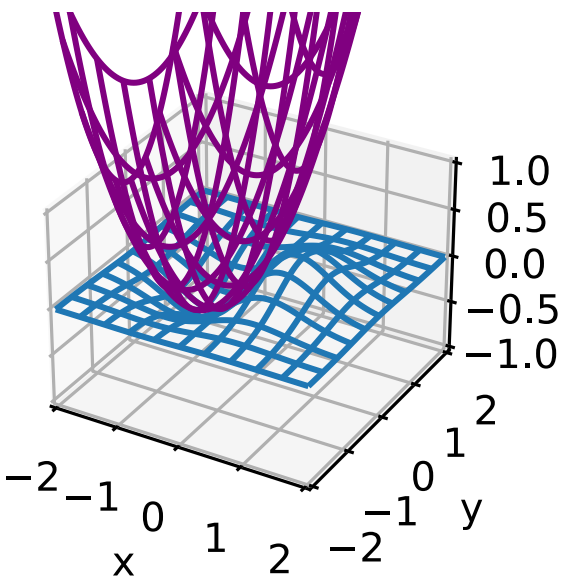

This forms the basis for Newton's Algorithm discussed in Section 12.3, where we perform numerical optimization iteratively finding the best fitting quadratic, and then exactly minimizing that quadratic.

### A.4.7 A Little Matrix Calculus

Derivatives of functions involving matrices turn out to be particularly nice. This section can become notationally heavy, so may be skipped in a first reading, but it is useful to know how derivatives of functions involving common matrix operations are often much cleaner than one might initially anticipate, particularly given how central matrix operations are to deep learning applications.

Let's begin with an example. Suppose that we have some fixed column vector $\boldsymbol{\beta}$, and we want to take the product function $f(\mathbf{x}) = \boldsymbol{\beta}^\top\mathbf{x}$, and understand how the dot product changes when we change $\mathbf{x}$.

A bit of notation that will be useful when working with matrix derivatives in ML is called the *denominator layout matrix derivative* where we assemble our partial derivatives into the shape of whatever vector, matrix, or tensor is in the denominator of the differential. In this case, we will write

$$\frac{df}{d\mathbf{x}} = \begin{bmatrix} \frac{df}{dx_1} \\ \vdots \\ \frac{df}{dx_n} \end{bmatrix}, \tag{A.34}$$

where we matched the shape of the column vector $\mathbf{x}$.

If we write out our function into components this is

$$f(\mathbf{x}) = \sum_{i=1}^{n} \beta_i x_i = \beta_1 x_1 + \cdots + \beta_n x_n. \tag{A.35}$$

If we now take the partial derivative with respect to say $\beta_1$, note that everything is zero but the first term, which is just $x_1$ multiplied by $\beta_1$, so we obtain that

$$\frac{df}{dx_1} = \beta_1, \tag{A.36}$$

or more generally that

$$\frac{df}{dx_i} = \beta_i. \tag{A.37}$$

We can now reassemble this into a matrix to see

$$\frac{df}{d\mathbf{x}} = \begin{bmatrix} \frac{df}{dx_1} \\ \vdots \\ \frac{df}{dx_n} \end{bmatrix} = \begin{bmatrix} \beta_1 \\ \vdots \\ \beta_n \end{bmatrix} = \boldsymbol{\beta}. \tag{A.38}$$

This illustrates a few factors about matrix calculus that we will often counter throughout this section:

- First, The computations will get rather involved.
- Second, The final results are much cleaner than the intermediate process, and will always look similar to the single variable case. In this case, note that $\frac{d}{dx}(bx) = b$ and $\frac{d}{d\mathbf{x}}(\boldsymbol{\beta}^\top \mathbf{x}) = \boldsymbol{\beta}$ are both similar.
- Third, transposes can often appear seemingly from nowhere. The core reason for this is the convention that we match the shape of the denominator, thus when we multiply matrices, we will need to take transposes to match back to the shape of the original term.

To keep building intuition, let's try a computation that is a little harder. Suppose that we have a column vector $\mathbf{x}$, and a square matrix $A$ and we want to compute

$$\frac{d}{d\mathbf{x}}(\mathbf{x}^\top A \mathbf{x}). \tag{A.39}$$

To drive towards easier to manipulate notation, let's consider this problem using Einstein notation. In this case we can write the function as

$$\mathbf{x}^\top A \mathbf{x} = x_i a_{ij} x_j. \tag{A.40}$$

To compute our derivative, we need to understand for every $k$, what is the value of

$$\frac{d}{dx_k}(\mathbf{x}^\top A \mathbf{x}) = \frac{d}{dx_k} x_i a_{ij} x_j. \tag{A.41}$$

By the product rule, this is

$$\frac{d}{dx_k} x_i a_{ij} x_j = \frac{dx_i}{dx_k} a_{ij} x_j + x_i a_{ij} \frac{dx_j}{dx_k}. \tag{A.42}$$

For a term like $\frac{dx_i}{dx_k}$, it is not hard to see that this is one when $i = k$ and zero otherwise. This means that every term where $i$ and $k$ are different vanish from this sum, so the only terms that remain in that first sum are the ones where $i = k$. The same reasoning holds for the second term where we need $j = k$. This gives

$$\frac{d}{dx_k} x_i a_{ij} x_j = a_{kj} x_j + x_i a_{ik}. \tag{A.43}$$

Now, the names of the indices in Einstein notation are arbitrary—the fact that $i$ and $j$ are different is immaterial to this computation at this point, so we can re-index so that they both use $i$ to see that

$$\frac{d}{dx_k} x_i a_{ij} x_j = a_{ki} x_i + x_i a_{ik} = (a_{ki} + a_{ik}) x_i. \tag{A.44}$$

Now, here is where we start to need some practice to go further. Let's try and identify this outcome in terms of matrix operations. $a_{ki} + a_{ik}$ is the $k, i$-th component of $\mathbf{A} + \mathbf{A}^\top$. This gives

$$\frac{d}{dx_k} x_i a_{ij} x_j = [\mathbf{A} + \mathbf{A}^\top]_{ki} x_i. \tag{A.45}$$

Similarly, this term is now the product of the matrix $\mathbf{A} + \mathbf{A}^\top$ by the vector $\mathbf{x}$, so we see that

$$\left[\frac{d}{d\mathbf{x}}(\mathbf{x}^\top A\mathbf{x})\right]_k = \frac{d}{dx_k} x_i a_{ij} x_j = [(\mathbf{A} + \mathbf{A}^\top)\mathbf{x}]_k. \tag{A.46}$$

Thus, we see that the $k$-th entry of the desired derivative from (A.39) is just the $k$-th entry of the vector on the right, and thus the two are the same. Thus yields

$$\frac{d}{d\mathbf{x}}(\mathbf{x}^\top A\mathbf{x}) = (\mathbf{A} + \mathbf{A}^\top)\mathbf{x}. \tag{A.47}$$

This required significantly more work than our last one, but the final result is small. More than that, consider the following computation for traditional single variable derivatives:

$$\frac{d}{dx}(xax) = \frac{dx}{dx} ax + xa\frac{dx}{dx} = (a + a)x. \tag{A.48}$$

Equivalently $\frac{d}{dx}(ax^2) = 2ax = (a + a)x$. Again, we get a result that looks rather like the single variable result but with a transpose tossed in.

At this point, the pattern should be looking rather suspicious, so let's try to figure out why. When we take matrix derivatives like this, let's first assume that the expression we get will be another matrix expression: an expression we can write it in terms of products and sums of matrices and their transposes. If such an expression exists, it will need to be true for all matrices. In particular, it will need to be true of $1 \times 1$ matrices, in which case the matrix product is just the product of the numbers, the matrix sum is just the sum, and the transpose does nothing at all! In other words, whatever expression we get *must* match the single variable expression. This means that, with some practice, one can often guess matrix derivatives just by knowing what the associated single variable expression must look like!

Let's try this out. Suppose that $\mathbf{X}$ is a $n \times m$ matrix, $\mathbf{U}$ is an $n \times r$ and $\mathbf{V}$ is an $r \times m$. Let's try to compute

$$\frac{d}{d\mathbf{V}}\|\mathbf{X} - \mathbf{U}\mathbf{V}\|_2^2 = \,? \tag{A.49}$$

This computation is important in an area called matrix factorization. For us, however, it is just a derivative to compute. Let's try to imagine what this would be for $1 \times 1$ matrices. In that case, we get the expression

$$\frac{d}{dv}(x - uv)^2 = -2(x - uv)u, \tag{A.50}$$

where, the derivative is rather standard. If we try to convert this back into a matrix expression we get

$$\frac{d}{d\mathbf{V}}\|\mathbf{X} - \mathbf{U}\mathbf{V}\|_2^2 = -2(\mathbf{X} - \mathbf{U}\mathbf{V})\mathbf{U}. \tag{A.51}$$

However, if we look at this it does not quite work. Recall that $\mathbf{X}$ is $n \times m$, as is $\mathbf{U}\mathbf{V}$, so the matrix $2(\mathbf{X} - \mathbf{U}\mathbf{V})$ is $n \times m$. On the other hand $\mathbf{U}$ is $n \times r$, and we cannot multiply a $n \times m$ and a $n \times r$ matrix since the dimensions do not match!

We want to get $\frac{d}{d\mathbf{V}}$, which is the same shape as $\mathbf{V}$, which is $r \times m$. So somehow we need to take a $n \times m$ matrix and a $n \times r$ matrix, multiply them together (perhaps with some transposes) to get a $r \times m$. We can do this by multiplying $U^\top$ by $(\mathbf{X} - \mathbf{U}\mathbf{V})$. Thus, we can guess the solution to (A.49) is

$$\frac{d}{d\mathbf{V}}\|\mathbf{X} - \mathbf{U}\mathbf{V}\|_2^2 = -2\mathbf{U}^\top(\mathbf{X} - \mathbf{U}\mathbf{V}). \tag{A.52}$$

To show that this works, we would be remiss to not provide a detailed computation. If we already believe that this rule-of-thumb works, feel free to skip past this derivation. To compute

$$\frac{d}{d\mathbf{V}}\|\mathbf{X} - \mathbf{U}\mathbf{V}\|_2^2, \tag{A.53}$$

we must find for every $a$, and $b$

$$\frac{d}{dv_{ab}}\|\mathbf{X} - \mathbf{U}\mathbf{V}\|_2^2 = \frac{d}{dv_{ab}}\sum_{i,j}\left(x_{ij} - \sum_k u_{ik}v_{kj}\right)^2. \tag{A.54}$$

Recalling that all entries of $\mathbf{X}$ and $\mathbf{U}$ are constants as far as $\frac{d}{dv_{ab}}$ is concerned, we may push the derivative inside the sum, and apply the chain rule to the square to get

$$\frac{d}{dv_{ab}}\|\mathbf{X} - \mathbf{U}\mathbf{V}\|_2^2 = \sum_{i,j} 2\left(x_{ij} - \sum_k u_{ik}v_{kj}\right)\left(-\sum_k u_{ik}\frac{dv_{kj}}{dv_{ab}}\right). \tag{A.55}$$

As in the previous derivation, we may note that $\frac{dv_{kj}}{dv_{ab}}$ is only non-zero if the $k = a$ and $j = b$. If either of those conditions do not hold, the term in the sum is zero, and we may freely discard it. We see that

$$\frac{d}{dv_{ab}}\|\mathbf{X} - \mathbf{U}\mathbf{V}\|_2^2 = -2\sum_i\left(x_{ib} - \sum_k u_{ik}v_{kb}\right)u_{ia}. \tag{A.56}$$

An important subtlety here is that the requirement that $k = a$ does not occur inside the inner sum since that $k$ is a dummy variable which we are summing over inside the inner term. For a notationally cleaner example, consider why

$$\frac{d}{dx_1}\left(\sum_i x_i\right)^2 = 2\left(\sum_i x_i\right). \tag{A.57}$$

From this point, we may start identifying components of the sum. First,

$$\sum_k u_{ik}v_{kb} = [\mathbf{UV}]_{ib}. \tag{A.58}$$

So the entire expression in the inside of the sum is

$$x_{ib} - \sum_k u_{ik}v_{kb} = [\mathbf{X} - \mathbf{UV}]_{ib}. \tag{A.59}$$

This means we may now write our derivative as

$$\frac{d}{dv_{ab}}\|\mathbf{X} - \mathbf{UV}\|_2^2 = -2\sum_i [\mathbf{X} - \mathbf{UV}]_{ib}u_{ia}. \tag{A.60}$$

We want this to look like the $a, b$ element of a matrix so we can use the technique as in the previous example to arrive at a matrix expression, which means that we need to exchange the order of the indices on $u_{ia}$. If we notice that $u_{ia} = [\mathbf{U}^\top]_{ai}$, we can then write

$$\frac{d}{dv_{ab}}\|\mathbf{X} - \mathbf{UV}\|_2^2 = -2\sum_i [\mathbf{U}^\top]_{ai}[\mathbf{X} - \mathbf{UV}]_{ib}. \tag{A.61}$$

This is a matrix product, and thus we can conclude that

$$\frac{d}{dv_{ab}}\|\mathbf{X} - \mathbf{UV}\|_2^2 = -2[\mathbf{U}^\top(\mathbf{X} - \mathbf{UV})]_{ab}. \tag{A.62}$$

and thus we may write the solution to (A.49)

$$\frac{d}{d\mathbf{V}}\|\mathbf{X} - \mathbf{UV}\|_2^2 = -2\mathbf{U}^\top(\mathbf{X} - \mathbf{UV}). \tag{A.63}$$

This matches the solution we guessed above!

It is reasonable to ask at this point, "Why can I not just write down matrix versions of all the calculus rules I have learned? It is clear this is still mechanical. Why do we not just get it over with!" And indeed there are such rules and (Petersen and Pedersen, 2008) provides an excellent summary. However, due to the plethora of ways matrix operations can be combined compared to single values, there are many more matrix derivative rules than single variable ones. It is often the case that it is best to work with the indices, or leave it up to automatic differentiation when appropriate.

### A.4.8 Summary

- In higher dimensions, we can define gradients which serve the same purpose as derivatives in one dimension. These allow us to see how a multi-variable function changes when we make an arbitrary small change to the inputs.

- The backpropagation algorithm can be seen to be a method of organizing the multi-variable chain rule to allow for the efficient computation of many partial derivatives.
- Matrix calculus allows us to write the derivatives of matrix expressions in concise ways.

### A.4.9 Exercises

1. Given a column vector $\boldsymbol{\beta}$, compute the derivatives of both $f(\mathbf{x}) = \boldsymbol{\beta}^\top \mathbf{x}$ and $g(\mathbf{x}) = \mathbf{x}^\top \boldsymbol{\beta}$. Why do you get the same answer?
2. Let $\mathbf{v}$ be an $n$ dimension vector. What is $\frac{\partial}{\partial \mathbf{v}} \|\mathbf{v}\|_2$?
3. Let $L(x, y) = \log(e^x + e^y)$. Compute the gradient. What is the sum of the components of the gradient?
4. Let $f(x, y) = x^2y + xy^2$. Show that the only critical point is $(0, 0)$. By considering $f(x, x)$, determine if $(0, 0)$ is a maximum, minimum, or neither.
5. Suppose that we are minimizing a function $f(\mathbf{x}) = g(\mathbf{x}) + h(\mathbf{x})$. How can we geometrically interpret the condition of $\nabla f = 0$ in terms of $g$ and $h$?

281

Discussions[281].

## A.5 Integral Calculus

Differentiation only makes up half of the content of a traditional calculus education. The other pillar, integration, starts out seeming a rather disjoint question, "What is the area underneath this curve?" While seemingly unrelated, integration is tightly intertwined with the differentiation via what is known as the *fundamental theorem of calculus*.

At the level of machine learning we discuss in this book, we will not need a deep understanding of integration. However, we will provide a brief introduction to lay the groundwork for any further applications we will encounter later on.

### A.5.1 Geometric Interpretation

Suppose that we have a function $f(x)$. For simplicity, let's assume that $f(x)$ is non-negative (never takes a value less than zero). What we want to try and understand is: what is the area contained between $f(x)$ and the $x$-axis?

```
%matplotlib inline
import torch
from IPython import display
from mpl_toolkits import mplot3d
from d2l import torch as d2l

x = torch.arange(-2, 2, 0.01)
f = torch.exp(-x**2)

d2l.set_figsize()
d2l.plt.plot(x, f, color='black')
d2l.plt.fill_between(x.tolist(), f.tolist())
d2l.plt.show()
```

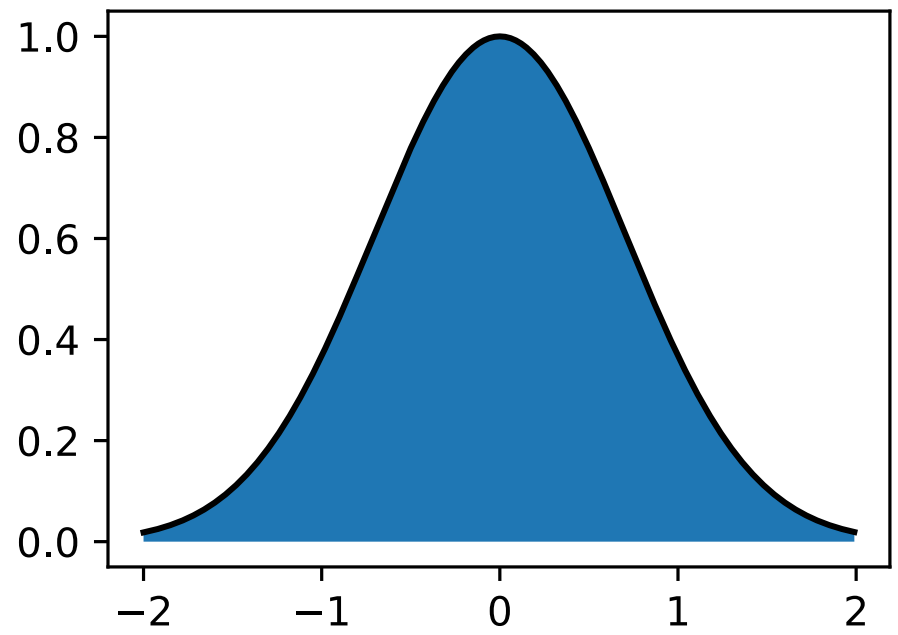

In most cases, this area will be infinite or undefined (consider the area under $f(x) = x^2$), so people will often talk about the area between a pair of ends, say $a$ and $b$.

```
x = torch.arange(-2, 2, 0.01)
f = torch.exp(-x**2)

d2l.set_figsize()
d2l.plt.plot(x, f, color='black')
d2l.plt.fill_between(x.tolist()[50:250], f.tolist()[50:250])
d2l.plt.show()
```

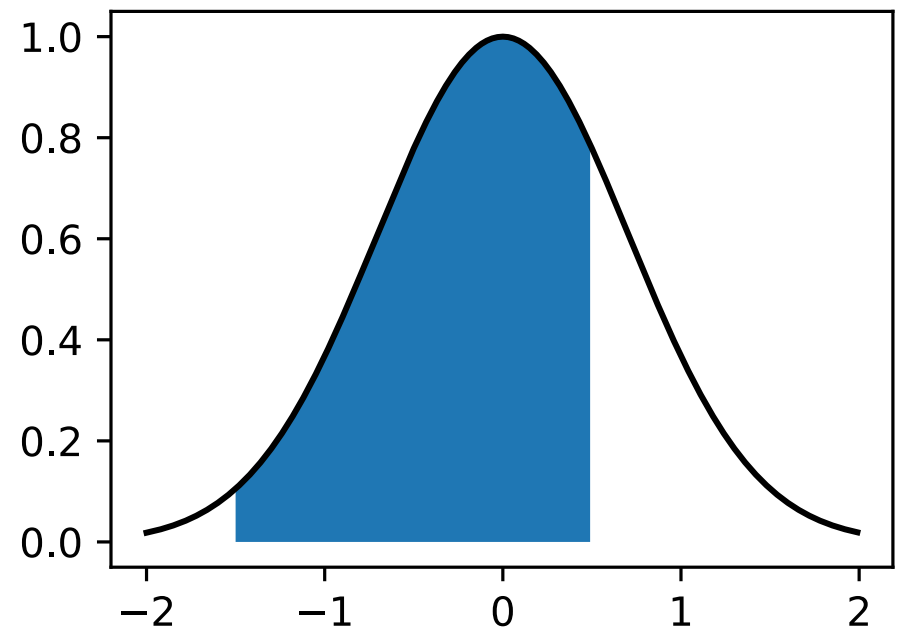

We will denote this area by the integral symbol below:

$$\text{Area}(\mathcal{A}) = \int_a^b f(x)\ dx. \tag{A.1}$$

The inner variable is a dummy variable, much like the index of a sum in a $\sum$, and so this can be equivalently written with any inner value we like:

$$\int_a^b f(x)\ dx = \int_a^b f(z)\ dz. \tag{A.2}$$

There is a traditional way to try and understand how we might try to approximate such integrals: we can imagine taking the region in-between $a$ and $b$ and chopping it into $N$ vertical slices. If $N$ is large, we can approximate the area of each slice by a rectangle, and then add up the areas to get the total area under the curve. Let's take a look at an example doing this in code. We will see how to get the true value in a later section.

```
epsilon = 0.05
a = 0
b = 2

x = torch.arange(a, b, epsilon)
f = x / (1 + x**2)

approx = torch.sum(epsilon*f)
true = torch.log(torch.tensor([5.])) / 2

d2l.set_figsize()
d2l.plt.bar(x, f, width=epsilon, align='edge')
d2l.plt.plot(x, f, color='black')
d2l.plt.ylim([0, 1])
d2l.plt.show()

f'approximation: {approx}, truth: {true}'
```

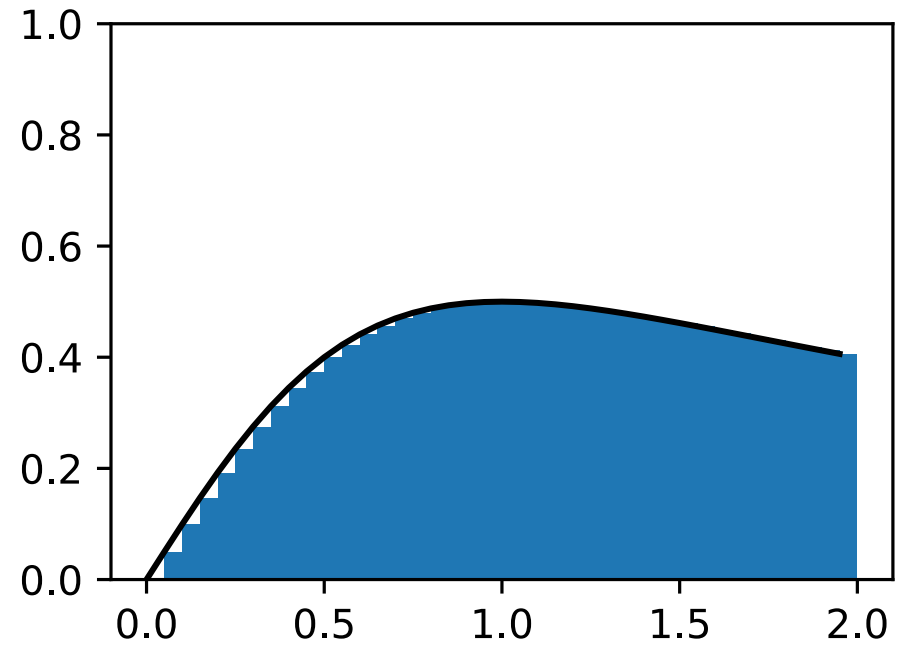

```
'approximation: 0.7944855690002441, truth: tensor([0.8047])'
```

The issue is that while it can be done numerically, we can do this approach analytically for only the simplest functions like

$$\int_a^b x \, dx. \tag{A.3}$$

Anything somewhat more complex like our example from the code above

$$\int_a^b \frac{x}{1+x^2} \, dx. \tag{A.4}$$

is beyond what we can solve with such a direct method.

We will instead take a different approach. We will work intuitively with the notion of the area, and learn the main computational tool used to find integrals: the *fundamental theorem of calculus*. This will be the basis for our study of integration.

### A.5.2 The Fundamental Theorem of Calculus

To dive deeper into the theory of integration, let's introduce a function

$$F(x) = \int_0^x f(y) dy. \tag{A.5}$$

This function measures the area between 0 and $x$ depending on how we change $x$. Notice that this is everything we need since

$$\int_a^b f(x) \, dx = F(b) - F(a). \tag{A.6}$$

This is a mathematical encoding of the fact that we can measure the area out to the far end-point and then subtract off the area to the near end point as indicated in Fig. A.1.

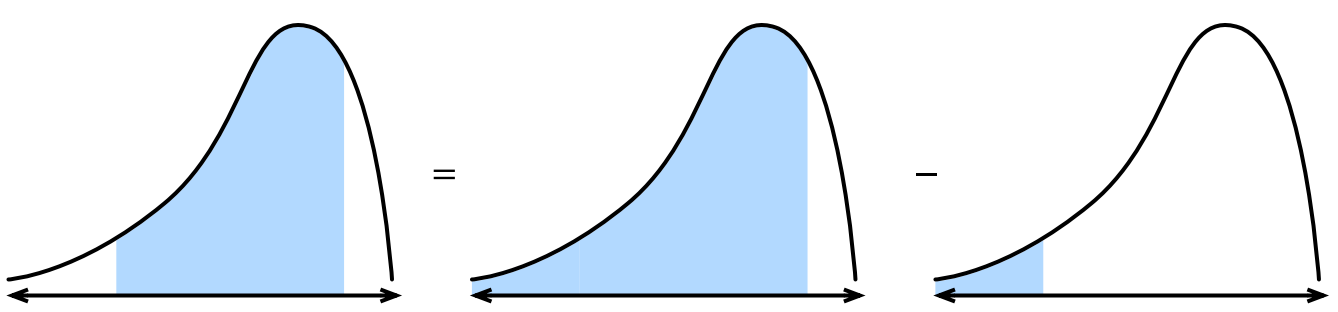

Fig. A.1 Visualizing why we may reduce the problem of computing the area under a curve between two points to computing the area to the left of a point.

Thus, we can figure out what the integral over any interval is by figuring out what $F(x)$ is.

To do so, let's consider an experiment. As we often do in calculus, let's imagine what happens when we shift the value by a tiny bit. From the comment above, we know that

$$F(x+\epsilon) - F(x) = \int_x^{x+\epsilon} f(y) \, dy. \tag{A.7}$$

This tells us that the function changes by the area under a tiny sliver of a function.

This is the point at which we make an approximation. If we look at a tiny sliver of area like this, it looks like this area is close to the rectangular area with height the value of $f(x)$ and the base width $\epsilon$. Indeed, one can show that as $\epsilon \rightarrow 0$ this approximation becomes better and better. Thus we can conclude:

$$F(x+\epsilon) - F(x) \approx \epsilon f(x). \tag{A.8}$$

However, we can now notice: this is exactly the pattern we expect if we were computing the derivative of $F$! Thus we see the following rather surprising fact:

$$\frac{dF}{dx}(x) = f(x). \tag{A.9}$$

This is the *fundamental theorem of calculus*. We may write it in expanded form as

$$\frac{d}{dx}\int_0^x f(y) \; dy = f(x). \tag{A.10}$$

It takes the concept of finding areas (*a priori* rather hard), and reduces it to a statement derivatives (something much more completely understood). One last comment that we must make is that this does not tell us exactly what $F(x)$ is. Indeed $F(x) + C$ for any $C$ has the same derivative. This is a fact-of-life in the theory of integration. Thankfully, notice that when working with definite integrals, the constants drop out, and thus are irrelevant to the outcome.

$$\int_a^b f(x) \; dx = (F(b) + C) - (F(a) + C) = F(b) - F(a). \tag{A.11}$$

This may seem like abstract non-sense, but let's take a moment to appreciate that it has given us a whole new perspective on computing integrals. Our goal is no-longer to do some sort of chop-and-sum process to try and recover the area, rather we need only find a function whose derivative is the function we have! This is incredible since we can now list many rather difficult integrals by just reversing the table from Section A.3.2. For instance, we know that the derivative of $x^n$ is $nx^{n-1}$. Thus, we can say using the fundamental theorem (A.10) that

$$\int_0^x ny^{n-1} \; dy = x^n - 0^n = x^n. \tag{A.12}$$

Similarly, we know that the derivative of $e^x$ is itself, so that means

$$\int_0^x e^x \; dx = e^x - e^0 = e^x - 1. \tag{A.13}$$

In this way, we can develop the entire theory of integration leveraging ideas from differential calculus freely. Every integration rule derives from this one fact.

### A.5.3 Change of Variables

Just as with differentiation, there are a number of rules which make the computation of integrals more tractable. In fact, every rule of differential calculus (like the product rule, sum rule, and chain rule) has a corresponding rule for integral calculus (integration by parts, linearity of integration, and the change of variables formula respectively). In this section, we will dive into what is arguably the most important from the list: the change of variables formula.

First, suppose that we have a function which is itself an integral:

$$F(x) = \int_0^x f(y) \, dy. \tag{A.14}$$

Let's suppose that we want to know how this function looks when we compose it with another to obtain $F(u(x))$. By the chain rule, we know

$$\frac{d}{dx} F(u(x)) = \frac{dF}{du}(u(x)) \cdot \frac{du}{dx}. \tag{A.15}$$

We can turn this into a statement about integration by using the fundamental theorem (A.10) as above. This gives

$$F(u(x)) - F(u(0)) = \int_0^x \frac{dF}{du}(u(y)) \cdot \frac{du}{dy} \, dy. \tag{A.16}$$

Recalling that $F$ is itself an integral gives that the left hand side may be rewritten to be

$$\int_{u(0)}^{u(x)} f(y) \, dy = \int_0^x \frac{dF}{du}(u(y)) \cdot \frac{du}{dy} \, dy. \tag{A.17}$$

Similarly, recalling that $F$ is an integral allows us to recognize that $\frac{dF}{dx} = f$ using the fundamental theorem (A.10), and thus we may conclude

$$\int_{u(0)}^{u(x)} f(y) \, dy = \int_0^x f(u(y)) \cdot \frac{du}{dy} \, dy. \tag{A.18}$$

This is the *change of variables* formula.

For a more intuitive derivation, consider what happens when we take an integral of $f(u(x))$ between $x$ and $x + \epsilon$. For a small $\epsilon$, this integral is approximately $\epsilon f(u(x))$, the area of the associated rectangle. Now, let's compare this with the integral of $f(y)$ from $u(x)$ to $u(x+\epsilon)$. We know that $u(x + \epsilon) \approx u(x) + \epsilon \frac{du}{dx}(x)$, so the area of this rectangle is approximately $\epsilon \frac{du}{dx}(x) f(u(x))$. Thus, to make the area of these two rectangles to agree, we need to multiply the first one by $\frac{du}{dx}(x)$ as is illustrated in Fig. A.2.

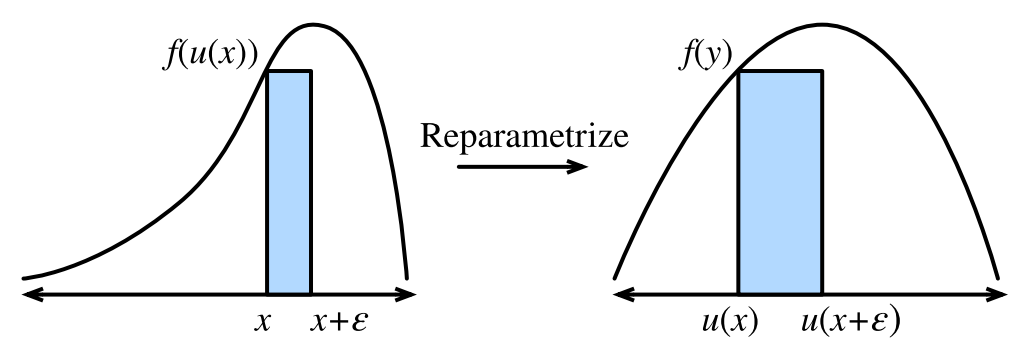

Fig. A.2 Visualizing the transformation of a single thin rectangle under the change of variables.

This tells us that

$$\int_x^{x+\epsilon} f(u(y)) \frac{du}{dy}(y) \, dy = \int_{u(x)}^{u(x+\epsilon)} f(y) \, dy. \tag{A.19}$$

This is the change of variables formula expressed for a single small rectangle.

If $u(x)$ and $f(x)$ are properly chosen, this can allow for the computation of incredibly complex integrals. For instance, if we even chose $f(y) = 1$ and $u(x) = e^{-x^2}$ (which means $\frac{du}{dx}(x) = -2xe^{-x^2}$), this can show for instance that

$$e^{-1} - 1 = \int_{e^{-0}}^{e^{-1}} 1 \; dy = -2 \int_0^1 ye^{-y^2} \; dy, \tag{A.20}$$

and thus by rearranging that

$$\int_0^1 ye^{-y^2} \; dy = \frac{1 - e^{-1}}{2}. \tag{A.21}$$

## A.5.4 A Comment on Sign Conventions

Keen-eyed readers will observe something strange about the computations above. Namely, computations like

$$\int_{e^{-0}}^{e^{-1}} 1 \; dy = e^{-1} - 1 < 0, \tag{A.22}$$

can produce negative numbers. When thinking about areas, it can be strange to see a negative value, and so it is worth digging into what the convention is.

Mathematicians take the notion of signed areas. This manifests itself in two ways. First, if we consider a function $f(x)$ which is sometimes less than zero, then the area will also be negative. So for instance

$$\int_0^1 (-1) \; dx = -1. \tag{A.23}$$

Similarly, integrals which progress from right to left, rather than left to right are also taken to be negative areas

$$\int_0^{-1} 1 \; dx = -1. \tag{A.24}$$

The standard area (from left to right of a positive function) is always positive. Anything obtained by flipping it (say flipping over the $x$-axis to get the integral of a negative number, or flipping over the $y$-axis to get an integral in the wrong order) will produce a negative area. And indeed, flipping twice will give a pair of negative signs that cancel out to have positive area

$$\int_0^{-1} (-1) \; dx = 1. \tag{A.25}$$

If this discussion sounds familiar, it is! In Section A.1 we discussed how the determinant represented the signed area in much the same way.

### A.5.5 Multiple Integrals

In some cases, we will need to work in higher dimensions. For instance, suppose that we have a function of two variables, like $f(x, y)$ and we want to know the volume under $f$ when $x$ ranges over $[a, b]$ and $y$ ranges over $[c, d]$.

```
# Construct grid and compute function
x, y = torch.meshgrid(torch.linspace(-2, 2, 101), torch.linspace(-2, 2, 101))
z = torch.exp(- x**2 - y**2)

# Plot function
ax = d2l.plt.figure().add_subplot(111, projection='3d')
ax.plot_wireframe(x, y, z)
d2l.plt.xlabel('x')
d2l.plt.ylabel('y')
d2l.plt.xticks([-2, -1, 0, 1, 2])
d2l.plt.yticks([-2, -1, 0, 1, 2])
d2l.set_figsize()
ax.set_xlim(-2, 2)
ax.set_ylim(-2, 2)
ax.set_zlim(0, 1)
ax.dist = 12
```

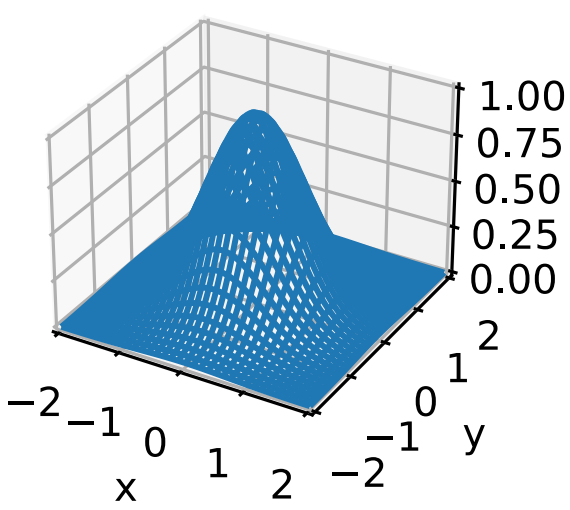

We write this as

$$\int_{[a,b]\times[c,d]} f(x, y) \; dx \; dy. \tag{A.26}$$

Suppose that we wish to compute this integral. My claim is that we can do this by iteratively computing first the integral in $x$ and then shifting to the integral in $y$, that is to say

$$\int_{[a,b]\times[c,d]} f(x, y) \; dx \; dy = \int_c^d \left(\int_a^b f(x, y) \; dx\right) \; dy. \tag{A.27}$$

Let's see why this is.

Consider the figure above where we have split the function into $\epsilon \times \epsilon$ squares which we will index with integer coordinates $i, j$. In this case, our integral is approximately

$$\sum_{i,j} \epsilon^2 f(\epsilon i, \epsilon j). \tag{A.28}$$

Once we discretize the problem, we may add up the values on these squares in whatever order we like, and not worry about changing the values. This is illustrated in Fig. A.3. In particular, we can say that

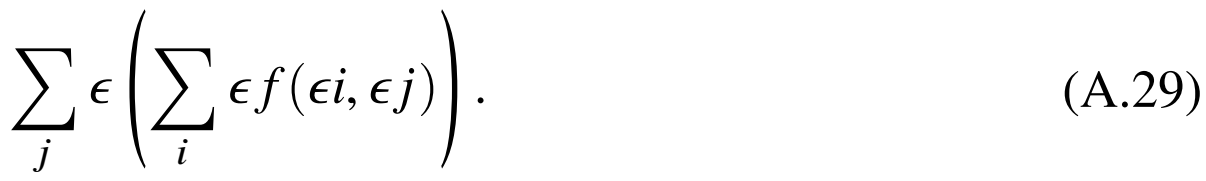

$$\sum_j \epsilon \left( \sum_i \epsilon f(\epsilon i, \epsilon j) \right). \tag{A.29}$$

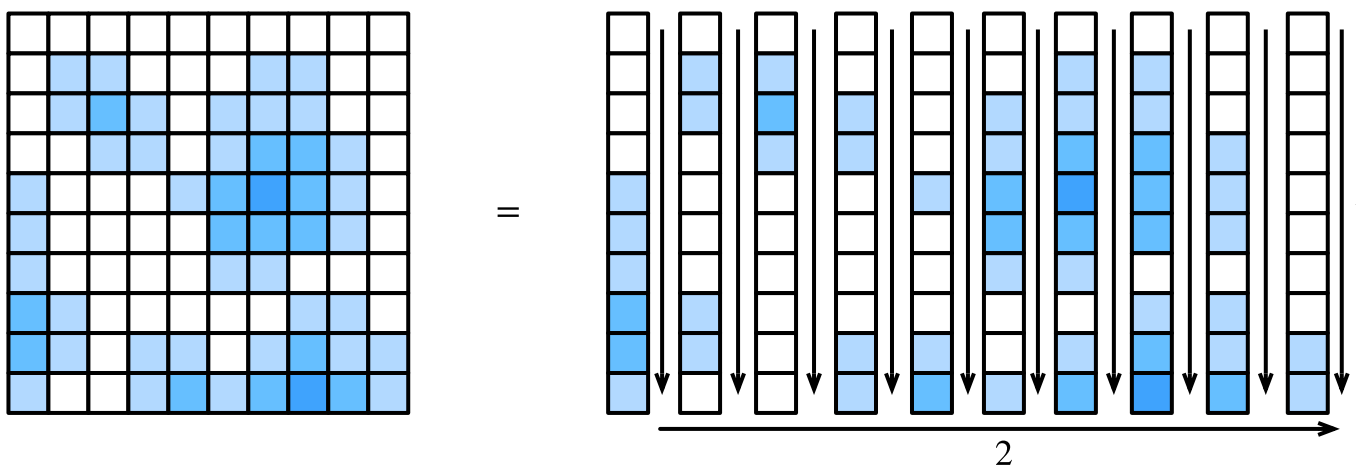

Fig. A.3 Illustrating how to decompose a sum over many squares as a sum over first the columns (1), then adding the column sums together (2).

The sum on the inside is precisely the discretization of the integral

$$G(\epsilon j) = \int_a^b f(x, \epsilon j) \; dx. \tag{A.30}$$

Finally, notice that if we combine these two expressions we get

$$\sum_j \epsilon G(\epsilon j) \approx \int_c^d G(y) \; dy = \int_{[a,b]\times[c,d]} f(x, y) \; dx \; dy. \tag{A.31}$$

Thus putting it all together, we have that

$$\int_{[a,b]\times[c,d]} f(x, y) \; dx \; dy = \int_c^d \left( \int_a^b f(x, y) \; dx \right) \; dy. \tag{A.32}$$

Notice that, once discretized, all we did was rearrange the order in which we added a list of numbers. This may make it seem like it is nothing, however this result (called *Fubini's Theorem*) is not always true! For the type of mathematics encountered when doing machine learning (continuous functions), there is no concern, however it is possible to create examples where it fails (for example the function $f(x, y) = xy(x^2 - y^2)/(x^2 + y^2)^3$ over the rectangle $[0, 2] \times [0, 1]$).

Note that the choice to do the integral in $x$ first, and then the integral in $y$ was arbitrary. We could have equally well chosen to do $y$ first and then $x$ to see

$$\int_{[a,b]\times[c,d]} f(x, y) \; dx \; dy = \int_a^b \left( \int_c^d f(x, y) \; dy \right) \; dx. \tag{A.33}$$

Often times, we will condense down to vector notation, and say that for $U = [a, b] \times [c, d]$ this is

$$\int_U f(\mathbf{x}) \; d\mathbf{x}. \tag{A.34}$$

## A.5.6 Change of Variables in Multiple Integrals

As with single variables in (A.18), the ability to change variables inside a higher dimensional integral is a key tool. Let's summarize the result without derivation.

We need a function that reparametrizes our domain of integration. We can take this to be $\phi : \mathbb{R}^n \to \mathbb{R}^n$, that is any function which takes in $n$ real variables and returns another $n$. To keep the expressions clean, we will assume that $\phi$ is *injective* which is to say it never folds over itself ($\phi(\mathbf{x}) = \phi(\mathbf{y}) \implies \mathbf{x} = \mathbf{y}$).

In this case, we can say that

$$\int_{\phi(U)} f(\mathbf{x})\ d\mathbf{x} = \int_U f(\phi(\mathbf{x})) \left|\det(D\phi(\mathbf{x}))\right|\ d\mathbf{x}. \tag{A.35}$$

where $D\phi$ is the *Jacobian* of $\phi$, which is the matrix of partial derivatives of $\boldsymbol{\phi} = (\phi_1(x_1, \ldots, x_n), \ldots, \phi_n(x_1, \ldots, x_n))$,

$$D\boldsymbol{\phi} = \begin{bmatrix} \frac{\partial \phi_1}{\partial x_1} & \cdots & \frac{\partial \phi_1}{\partial x_n} \\ \vdots & \ddots & \vdots \\ \frac{\partial \phi_n}{\partial x_1} & \cdots & \frac{\partial \phi_n}{\partial x_n} \end{bmatrix}. \tag{A.36}$$

Looking closely, we see that this is similar to the single variable chain rule (A.18), except we have replaced the term $\frac{du}{dx}(x)$ with $\left|\det(D\phi(\mathbf{x}))\right|$. Let's see how we can to interpret this term. Recall that the $\frac{du}{dx}(x)$ term existed to say how much we stretched our $x$-axis by applying $u$. The same process in higher dimensions is to determine how much we stretch the area (or volume, or hyper-volume) of a little square (or little *hyper-cube*) by applying $\boldsymbol{\phi}$. If $\boldsymbol{\phi}$ was the multiplication by a matrix, then we know how the determinant already gives the answer.

With some work, one can show that the *Jacobian* provides the best approximation to a multivariable function $\boldsymbol{\phi}$ at a point by a matrix in the same way we could approximate by lines or planes with derivatives and gradients. Thus the determinant of the Jacobian exactly mirrors the scaling factor we identified in one dimension.

It takes some work to fill in the details to this, so do not worry if they are not clear now. Let's see at least one example we will make use of later on. Consider the integral

$$\int_{-\infty}^{\infty} \int_{-\infty}^{\infty} e^{-x^2-y^2}\ dx\ dy. \tag{A.37}$$

Playing with this integral directly will get us no-where, but if we change variables, we can make significant progress. If we let $\boldsymbol{\phi}(r, \theta) = (r\cos(\theta), r\sin(\theta))$ (which is to say that $x = r\cos(\theta)$, $y = r\sin(\theta)$), then we can apply the change of variable formula to see that this is the same thing as

$$\int_0^{\infty} \int_0^{2\pi} e^{-r^2} \left|\det(D\text{Œ}(\mathbf{x}))\right|\ d\theta\ dr, \tag{A.38}$$

where

$$\left|\det(D\text{Œ}(\mathbf{x}))\right| = \left|\det \begin{bmatrix} \cos(\theta) & -r\sin(\theta) \\ \sin(\theta) & r\cos(\theta) \end{bmatrix}\right| = r(\cos^2(\theta) + \sin^2(\theta)) = r. \tag{A.39}$$

Thus, the integral is

$$\int_0^\infty \int_0^{2\pi} re^{-r^2} \; d\theta \; dr = 2\pi \int_0^\infty re^{-r^2} \; dr = \pi, \tag{A.40}$$

where the final equality follows by the same computation that we used in section Section A.5.3.

We will meet this integral again when we study continuous random variables in Section A.6.

### A.5.7 Summary

- The theory of integration allows us to answer questions about areas or volumes.
- The fundamental theorem of calculus allows us to leverage knowledge about derivatives to compute areas via the observation that the derivative of the area up to some point is given by the value of the function being integrated.
- Integrals in higher dimensions can be computed by iterating single variable integrals.

### A.5.8 Exercises

1. What is $\int_1^2 \frac{1}{x} \; dx$?
2. Use the change of variables formula to integrate $\int_0^{\sqrt{\pi}} x \sin(x^2) \; dx$.
3. What is $\int_{[0,1]^2} xy \; dx \; dy$?
4. Use the change of variables formula to compute $\int_0^2 \int_0^1 xy(x^2 - y^2)/(x^2 + y^2)^3 \; dy \; dx$ and $\int_0^1 \int_0^2 f(x, y) = xy(x^2 - y^2)/(x^2 + y^2)^3 \; dx \; dy$ to see they are different.

Discussions[282].

## A.6 Random Variables

In Section 2.6 we saw the basics of how to work with discrete random variables, which in our case refer to those random variables which take either a finite set of possible values, or the integers. In this section, we develop the theory of *continuous random variables*, which are random variables which can take on any real value.

## A.6.1 Continuous Random Variables

Continuous random variables are a significantly more subtle topic than discrete random variables. A fair analogy to make is that the technical jump is comparable to the jump between adding lists of numbers and integrating functions. As such, we will need to take some time to develop the theory.

### From Discrete to Continuous

To understand the additional technical challenges encountered when working with continuous random variables, let's perform a thought experiment. Suppose that we are throwing a dart at the dart board, and we want to know the probability that it hits exactly 2cm from the center of the board.

To start with, we imagine measuring a single digit of accuracy, that is to say with bins for $0$cm, $1$cm, $2$cm, and so on. We throw say $100$ darts at the dart board, and if $20$ of them fall into the bin for 2cm we conclude that $20\%$ of the darts we throw hit the board 2cm away from the center.

However, when we look closer, this does not match our question! We wanted exact equality, whereas these bins hold all that fell between say $1.5$cm and $2.5$cm.

Undeterred, we continue further. We measure even more precisely, say $1.9$cm, $2.0$cm, $2.1$cm, and now see that perhaps 3 of the $100$ darts hit the board in the $2.0$cm bucket. Thus we conclude the probability is $3\%$.

However, this does not solve anything! We have just pushed the issue down one digit further. Let's abstract a bit. Imagine we know the probability that the first $k$ digits match with $2.00000\ldots$ and we want to know the probability it matches for the first $k+1$ digits. It is fairly reasonable to assume that the $k+1^{\mathrm{th}}$ digit is essentially a random choice from the set $\{0, 1, 2, \ldots, 9\}$. At least, we cannot conceive of a physically meaningful process which would force the number of micrometers away form the center to prefer to end in a 7 vs a 3.

What this means is that in essence each additional digit of accuracy we require should decrease probability of matching by a factor of $10$. Or put another way, we would expect that

$$P(\text{distance is } 2.00\ldots, \text{ to } k \text{ digits}) \approx p \cdot 10^{-k}. \tag{A.1}$$

The value $p$ essentially encodes what happens with the first few digits, and the $10^{-k}$ handles the rest.

Notice that if we know the position accurate to $k=4$ digits after the decimal, that means we know the value falls within the interval say $[1.99995, 2.00005]$ which is an interval of length $2.00005 - 1.99995 = 10^{-4}$. Thus, if we call the length of this interval $\epsilon$, we can say

$$P(\text{distance is in an } \epsilon\text{-sized interval around } 2) \approx \epsilon \cdot p. \tag{A.2}$$

Let's take this one final step further. We have been thinking about the point 2 the entire time,

but never thinking about other points. Nothing is different there fundamentally, but it is the case that the value $p$ will likely be different. We would at least hope that a dart thrower was more likely to hit a point near the center, like 2cm rather than 20cm. Thus, the value $p$ is not fixed, but rather should depend on the point $x$. This tells us that we should expect

$$P(\text{distance is in an } \epsilon\text{-sized interval around } x) \approx \epsilon \cdot p(x). \tag{A.3}$$

Indeed, (A.3) precisely defines the *probability density function*. It is a function $p(x)$ which encodes the relative probability of hitting near one point vs. another. Let's visualize what such a function might look like.

```
%matplotlib inline
import torch
from IPython import display
from d2l import torch as d2l

torch.pi = torch.acos(torch.zeros(1)).item() * 2  # Define pi in torch

# Plot the probability density function for some random variable
x = torch.arange(-5, 5, 0.01)
p = 0.2*torch.exp(-(x - 3)**2 / 2)/torch.sqrt(2 * torch.tensor(torch.pi)) + \
    0.8*torch.exp(-(x + 1)**2 / 2)/torch.sqrt(2 * torch.tensor(torch.pi))

d2l.plot(x, p, 'x', 'Density')
```

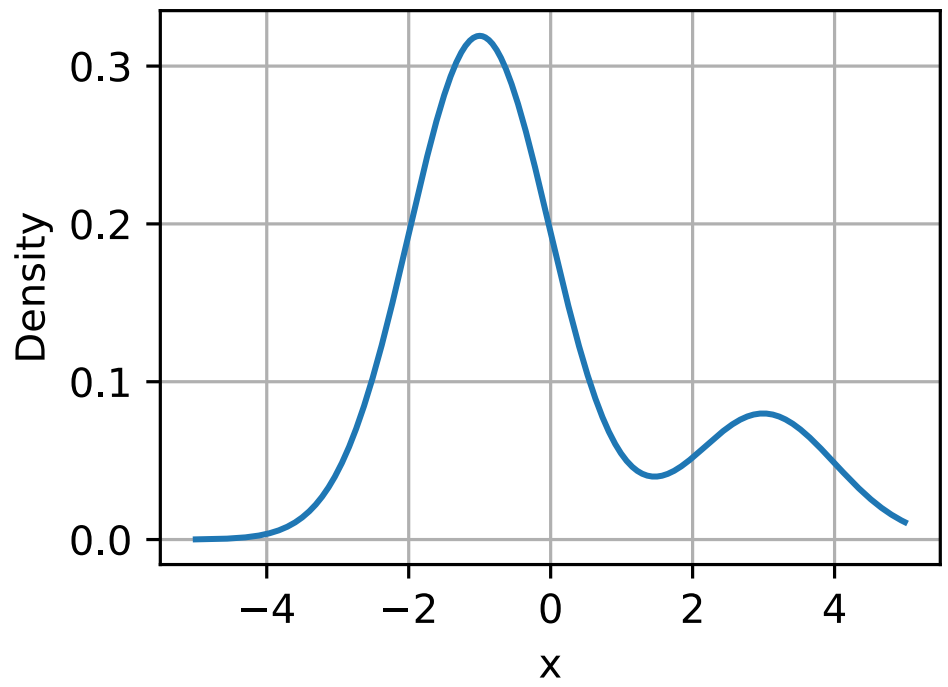

The locations where the function value is large indicates regions where we are more likely to find the random value. The low portions are areas where we are unlikely to find the random value.

## Probability Density Functions

Let's now investigate this further. We have already seen what a probability density function is intuitively for a random variable $X$, namely the density function is a function $p(x)$ so that

$$P(X \text{ is in an } \epsilon\text{-sized interval around } x) \approx \epsilon \cdot p(x). \tag{A.4}$$

But what does this imply for the properties of $p(x)$?

First, probabilities are never negative, thus we should expect that $p(x) \geq 0$ as well.

Second, let's imagine that we slice up the $\mathbb{R}$ into an infinite number of slices which are $\epsilon$ wide, say with slices $(\epsilon \cdot i, \epsilon \cdot (i+1)]$. For each of these, we know from (A.4) the probability is approximately

$$P(X \text{ is in an } \epsilon\text{-sized interval around } x) \approx \epsilon \cdot p(\epsilon \cdot i), \tag{A.5}$$

so summed over all of them it should be

$$P(X \in \mathbb{R}) \approx \sum_i \epsilon \cdot p(\epsilon \cdot i). \tag{A.6}$$

This is nothing more than the approximation of an integral discussed in Section A.5, thus we can say that

$$P(X \in \mathbb{R}) = \int_{-\infty}^{\infty} p(x) \; dx. \tag{A.7}$$

We know that $P(X \in \mathbb{R}) = 1$, since the random variable must take on *some* number, we can conclude that for any density

$$\int_{-\infty}^{\infty} p(x) \; dx = 1. \tag{A.8}$$

Indeed, digging into this further shows that for any $a$, and $b$, we see that

$$P(X \in (a, b]) = \int_{a}^{b} p(x) \; dx. \tag{A.9}$$

We may approximate this in code by using the same discrete approximation methods as before. In this case we can approximate the probability of falling in the blue region.

```
# Approximate probability using numerical integration
epsilon = 0.01
x = torch.arange(-5, 5, 0.01)
p = 0.2*torch.exp(-(x - 3)**2 / 2) / torch.sqrt(2 * torch.tensor(torch.pi)) +\
    0.8*torch.exp(-(x + 1)**2 / 2) / torch.sqrt(2 * torch.tensor(torch.pi))

d2l.set_figsize()
d2l.plt.plot(x, p, color='black')
d2l.plt.fill_between(x.tolist()[300:800], p.tolist()[300:800])
d2l.plt.show()

f'approximate Probability: {torch.sum(epsilon*p[300:800])}'
```

```
'approximate Probability: 0.773617148399353'
```

It turns out that these two properties describe exactly the space of possible probability density functions (or *p.d.f.*'s for the commonly encountered abbreviation). They are non-negative functions $p(x) \geq 0$ such that

$$\int_{-\infty}^{\infty} p(x) \; dx = 1. \tag{A.10}$$

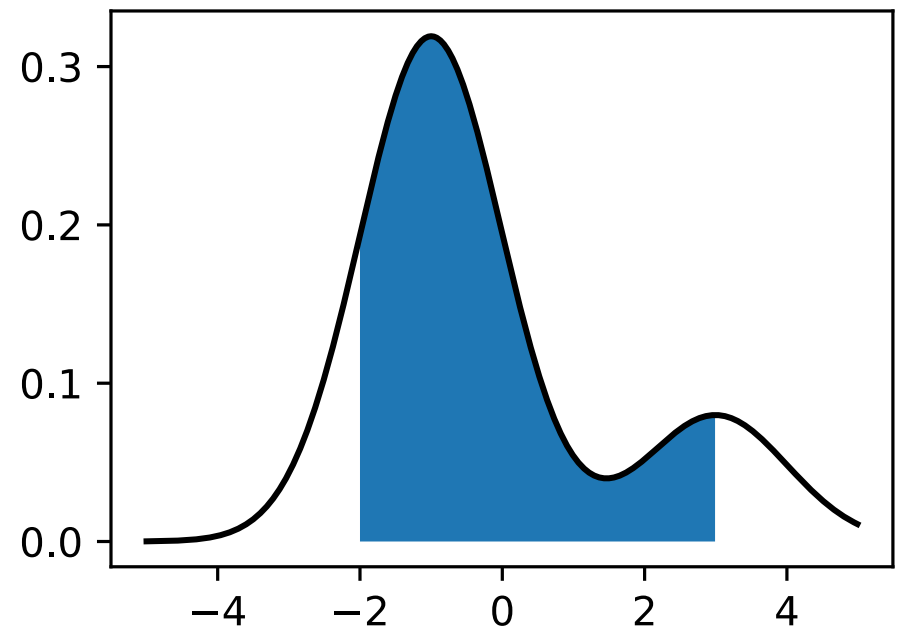

We interpret this function by using integration to obtain the probability our random variable is in a specific interval:

$$P(X \in (a, b]) = \int_a^b p(x) \; dx. \tag{A.11}$$

In Section A.8 we will see a number of common distributions, but let's continue working in the abstract.

## Cumulative Distribution Functions

In the previous section, we saw the notion of the p.d.f. In practice, this is a commonly encountered method to discuss continuous random variables, but it has one significant pitfall: that the values of the p.d.f. are not themselves probabilities, but rather a function that we must integrate to yield probabilities. There is nothing wrong with a density being larger than $10$, as long as it is not larger than $10$ for more than an interval of length $1/10$. This can be counter-intuitive, so people often also think in terms of the *cumulative distribution function*, or c.d.f., which *is* a probability.

In particular, by using (A.11), we define the c.d.f. for a random variable $X$ with density $p(x)$ by

$$F(x) = \int_{-\infty}^{x} p(x) \; dx = P(X \leq x). \tag{A.12}$$

Let's observe a few properties.

- $F(x) \rightarrow 0$ as $x \rightarrow -\infty$.
- $F(x) \rightarrow 1$ as $x \rightarrow \infty$.
- $F(x)$ is non-decreasing ($y > x \implies F(y) \geq F(x)$).
- $F(x)$ is continuous (has no jumps) if $X$ is a continuous random variable.

With the fourth bullet point, note that this would not be true if $X$ were discrete, say taking the values 0 and 1 both with probability $1/2$. In that case

$$F(x) = \begin{cases} 0 & x < 0, \\ \frac{1}{2} & x < 1, \\ 1 & x \geq 1. \end{cases} \tag{A.13}$$

In this example, we see one of the benefits of working with the c.d.f., the ability to deal with continuous or discrete random variables in the same framework, or indeed mixtures of the two (flip a coin: if heads return the roll of a die, if tails return the distance of a dart throw from the center of a dart board).

## Means

Suppose that we are dealing with a random variables $X$. The distribution itself can be hard to interpret. It is often useful to be able to summarize the behavior of a random variable concisely. Numbers that help us capture the behavior of a random variable are called *summary statistics*. The most commonly encountered ones are the *mean*, the *variance*, and the *standard deviation*.

The *mean* encodes the average value of a random variable. If we have a discrete random variable $X$, which takes the values $x_i$ with probabilities $p_i$, then the mean is given by the weighted average: sum the values times the probability that the random variable takes on that value:

$$\mu_X = E[X] = \sum_i x_i p_i. \tag{A.14}$$

The way we should interpret the mean (albeit with caution) is that it tells us essentially where the random variable tends to be located.

As a minimalistic example that we will examine throughout this section, let's take $X$ to be the random variable which takes the value $a - 2$ with probability $p$, $a + 2$ with probability $p$ and $a$ with probability $1 - 2p$. We can compute using (A.14) that, for any possible choice of $a$ and $p$, the mean is

$$\mu_X = E[X] = \sum_i x_i p_i = (a-2)p + a(1-2p) + (a+2)p = a. \tag{A.15}$$

Thus we see that the mean is $a$. This matches the intuition since $a$ is the location around which we centered our random variable.

Because they are helpful, let's summarize a few properties.

- For any random variable $X$ and numbers $a$ and $b$, we have that $\mu_{aX+b} = a\mu_X + b$.
- If we have two random variables $X$ and $Y$, we have $\mu_{X+Y} = \mu_X + \mu_Y$.

Means are useful for understanding the average behavior of a random variable, however the mean is not sufficient to even have a full intuitive understanding. Making a profit of $\$10 \pm \$1$ per sale is very different from making $\$10 \pm \$15$ per sale despite having the same average value. The second one has a much larger degree of fluctuation, and thus represents a much larger risk. Thus, to understand the behavior of a random variable, we will need at minimum one more measure: some measure of how widely a random variable fluctuates.

## Variances

This leads us to consider the *variance* of a random variable. This is a quantitative measure of how far a random variable deviates from the mean. Consider the expression $X - \mu_X$. This is the deviation of the random variable from its mean. This value can be positive or negative, so we need to do something to make it positive so that we are measuring the magnitude of the deviation.

A reasonable thing to try is to look at $|X - \mu_X|$, and indeed this leads to a useful quantity called the *mean absolute deviation*, however due to connections with other areas of mathematics and statistics, people often use a different solution.

In particular, they look at $(X - \mu_X)^2$. If we look at the typical size of this quantity by taking the mean, we arrive at the variance

$$\sigma_X^2 = \mathrm{Var}(X) = E\left[(X - \mu_X)^2\right] = E[X^2] - \mu_X^2. \tag{A.16}$$

The last equality in (A.16) holds by expanding out the definition in the middle, and applying the properties of expectation.

Let's look at our example where $X$ is the random variable which takes the value $a - 2$ with probability $p$, $a + 2$ with probability $p$ and $a$ with probability $1 - 2p$. In this case $\mu_X = a$, so all we need to compute is $E\left[X^2\right]$. This can readily be done:

$$E\left[X^2\right] = (a-2)^2 p + a^2(1-2p) + (a+2)^2 p = a^2 + 8p. \tag{A.17}$$

Thus, we see that by (A.16) our variance is

$$\sigma_X^2 = \mathrm{Var}(X) = E[X^2] - \mu_X^2 = a^2 + 8p - a^2 = 8p. \tag{A.18}$$

This result again makes sense. The largest $p$ can be is $1/2$ which corresponds to picking $a - 2$ or $a + 2$ with a coin flip. The variance of this being $4$ corresponds to the fact that both $a - 2$ and $a + 2$ are $2$ units away from the mean, and $2^2 = 4$. On the other end of the spectrum, if $p = 0$, this random variable always takes the value $0$ and so it has no variance at all.

We will list a few properties of variance below:

- For any random variable $X$, $\mathrm{Var}(X) \geq 0$, with $\mathrm{Var}(X) = 0$ if and only if $X$ is a constant.
- For any random variable $X$ and numbers $a$ and $b$, we have that $\mathrm{Var}(aX + b) = a^2\mathrm{Var}(X)$.
- If we have two *independent* random variables $X$ and $Y$, we have $\mathrm{Var}(X + Y) = \mathrm{Var}(X) + \mathrm{Var}(Y)$.

When interpreting these values, there can be a bit of a hiccup. In particular, let's try imagining what happens if we keep track of units through this computation. Suppose that we are working with the star rating assigned to a product on the web page. Then $a$, $a-2$, and $a+2$ are all measured in units of stars. Similarly, the mean $\mu_X$ is then also measured in stars (being a weighted average). However, if we get to the variance, we immediately encounter an issue, which is we want to look at $(X-\mu_X)^2$, which is in units of *squared stars*. This means that the variance itself is not comparable to the original measurements. To make it interpretable, we will need to return to our original units.

## Standard Deviations

This summary statistics can always be deduced from the variance by taking the square root! Thus we define the *standard deviation* to be

$$\sigma_X = \sqrt{\mathrm{Var}(X)}. \tag{A.19}$$

In our example, this means we now have the standard deviation is $\sigma_X = 2\sqrt{2p}$. If we are dealing with units of stars for our review example, $\sigma_X$ is again in units of stars.

The properties we had for the variance can be restated for the standard deviation.

- For any random variable $X$, $\sigma_X \geq 0$.
- For any random variable $X$ and numbers $a$ and $b$, we have that $\sigma_{aX+b} = |a|\sigma_X$
- If we have two *independent* random variables $X$ and $Y$, we have $\sigma_{X+Y} = \sqrt{\sigma_X^2 + \sigma_Y^2}$.

It is natural at this moment to ask, "If the standard deviation is in the units of our original random variable, does it represent something we can draw with regards to that random variable?" The answer is a resounding yes! Indeed much like the mean told us the typical location of our random variable, the standard deviation gives the typical range of variation of that random variable. We can make this rigorous with what is known as Chebyshev's inequality:

$$P\left(X \notin [\mu_X - \alpha\sigma_X, \mu_X + \alpha\sigma_X]\right) \leq \frac{1}{\alpha^2}. \tag{A.20}$$

Or to state it verbally in the case of $\alpha = 10$, $99\%$ of the samples from any random variable fall within 10 standard deviations of the mean. This gives an immediate interpretation to our standard summary statistics.

To see how this statement is rather subtle, let's take a look at our running example again where $X$ is the random variable which takes the value $a-2$ with probability $p$, $a+2$ with probability $p$ and $a$ with probability $1-2p$. We saw that the mean was $a$ and the standard deviation was $2\sqrt{2p}$. This means, if we take Chebyshev's inequality (A.20) with $\alpha = 2$, we see that the expression is

$$P\left(X \notin [a - 4\sqrt{2p}, a + 4\sqrt{2p}]\right) \leq \frac{1}{4}. \tag{A.21}$$

This means that $75\%$ of the time, this random variable will fall within this interval for any value of $p$. Now, notice that as $p \to 0$, this interval also converges to the single point $a$. But we know that our random variable takes the values $a-2$, $a$, and $a+2$ only so eventually we can be certain $a-2$ and $a+2$ will fall outside the interval! The question is, at what $p$ does that happen. So we want to solve: for what $p$ does $a+4\sqrt{2p}=a+2$, which is solved when $p=1/8$, which is *exactly* the first $p$ where it could possibly happen without violating our claim that no more than $1/4$ of samples from the distribution would fall outside the interval ($1/8$ to the left, and $1/8$ to the right).

Let's visualize this. We will show the probability of getting the three values as three vertical bars with height proportional to the probability. The interval will be drawn as a horizontal line in the middle. The first plot shows what happens for $p>1/8$ where the interval safely contains all points.

```
# Define a helper to plot these figures
def plot_chebyshev(a, p):
    d2l.set_figsize()
    d2l.plt.stem([a-2, a, a+2], [p, 1-2*p, p], use_line_collection=True)
    d2l.plt.xlim([-4, 4])
    d2l.plt.xlabel('x')
    d2l.plt.ylabel('p.m.f.')

    d2l.plt.hlines(0.5, a - 4 * torch.sqrt(2 * p),
                   a + 4 * torch.sqrt(2 * p), 'black', lw=4)
    d2l.plt.vlines(a - 4 * torch.sqrt(2 * p), 0.53, 0.47, 'black', lw=1)
    d2l.plt.vlines(a + 4 * torch.sqrt(2 * p), 0.53, 0.47, 'black', lw=1)
    d2l.plt.title(f'p = {p:.3f}')

    d2l.plt.show()

# Plot interval when p > 1/8
plot_chebyshev(0.0, torch.tensor(0.2))
```

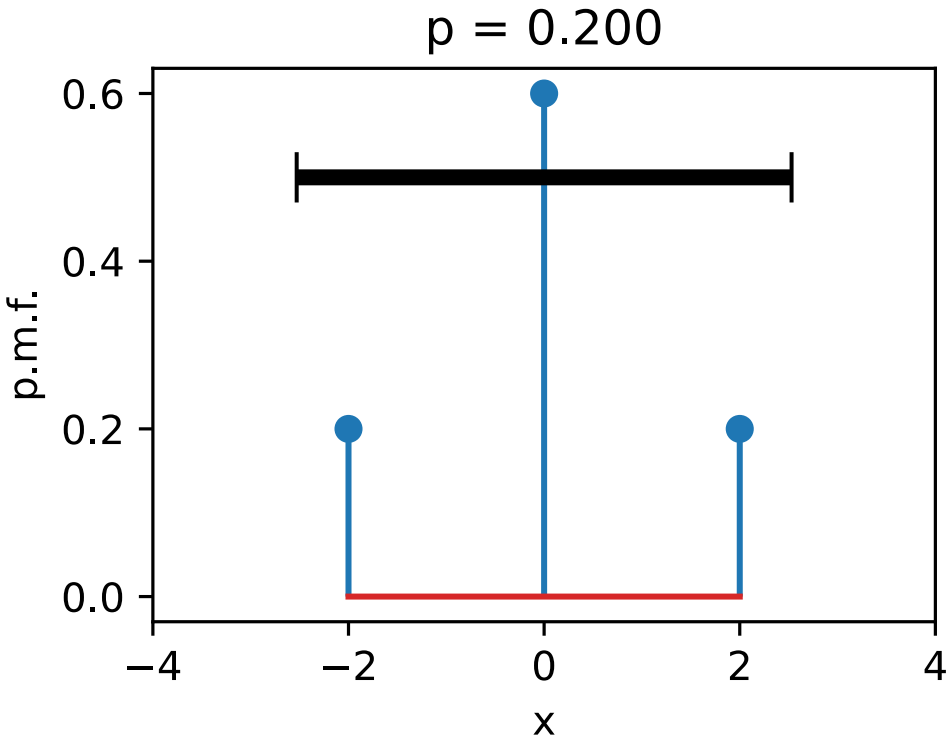

The second shows that at $p=1/8$, the interval exactly touches the two points. This shows that the inequality is *sharp*, since no smaller interval could be taken while keeping the inequality true.

```
# Plot interval when p = 1/8
plot_chebyshev(0.0, torch.tensor(0.125))
```

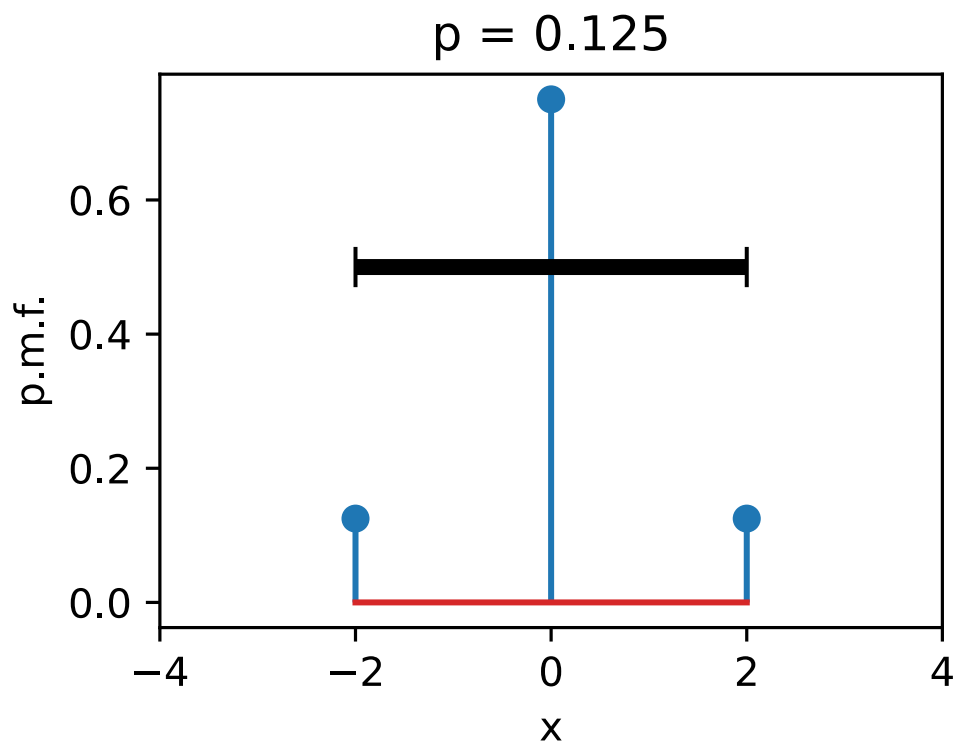

The third shows that for $p < 1/8$ the interval only contains the center. This does not invalidate the inequality since we only needed to ensure that no more than $1/4$ of the probability falls outside the interval, which means that once $p < 1/8$, the two points at $a - 2$ and $a + 2$ can be discarded.

```
# Plot interval when p < 1/8
plot_chebyshev(0.0, torch.tensor(0.05))
```

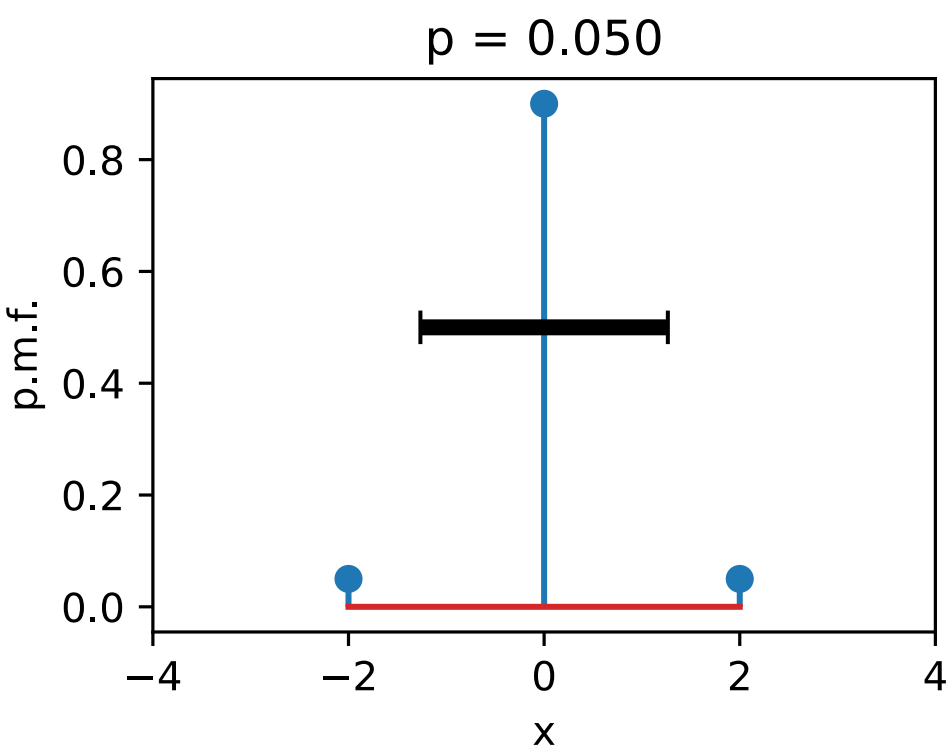

## Means and Variances in the Continuum

This has all been in terms of discrete random variables, but the case of continuous random variables is similar. To intuitively understand how this works, imagine that we split the real number line into intervals of length $\epsilon$ given by $(\epsilon i, \epsilon(i+1)]$. Once we do this, our continuous random variable has been made discrete and we can use (A.14) say that

$$
\begin{aligned}
\mu_X &\approx \sum_i (\epsilon i) P(X \in (\epsilon i, \epsilon(i+1)]) \\
&\approx \sum_i (\epsilon i) p_X(\epsilon i) \epsilon,
\end{aligned}
\tag{A.22}
$$

where $p_X$ is the density of $X$. This is an approximation to the integral of $xp_X(x)$, so we can conclude that

$$\mu_X = \int_{-\infty}^{\infty} xp_X(x) \; dx. \tag{A.23}$$

Similarly, using (A.16) the variance can be written as

$$\sigma^2_X = E[X^2] - \mu^2_X = \int_{-\infty}^{\infty} x^2p_X(x) \; dx - \left(\int_{-\infty}^{\infty} xp_X(x) \; dx\right)^2. \tag{A.24}$$

Everything stated above about the mean, the variance, and the standard deviation still applies in this case. For instance, if we consider the random variable with density

$$p(x) = \begin{cases} 1 & x \in [0,1], \\ 0 & \text{otherwise}. \end{cases} \tag{A.25}$$

we can compute

$$\mu_X = \int_{-\infty}^{\infty} xp(x) \; dx = \int_0^1 x \; dx = \frac{1}{2}. \tag{A.26}$$

and

$$\sigma_X^2 = \int_{-\infty}^{\infty} x^2p(x) \; dx - \left(\frac{1}{2}\right)^2 = \frac{1}{3} - \frac{1}{4} = \frac{1}{12}. \tag{A.27}$$

As a warning, let's examine one more example, known as the *Cauchy distribution*. This is the distribution with p.d.f. given by

$$p(x) = \frac{1}{1+x^2}. \tag{A.28}$$

```
# Plot the Cauchy distribution p.d.f.
x = torch.arange(-5, 5, 0.01)
p = 1 / (1 + x**2)

d2l.plot(x, p, 'x', 'p.d.f.')
```

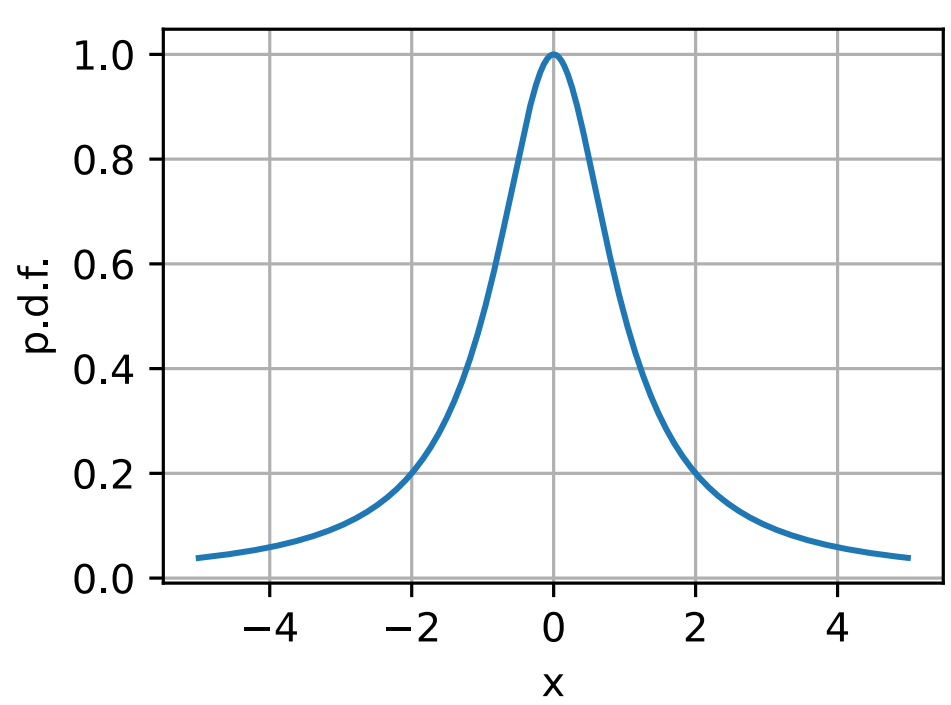

This function looks innocent, and indeed consulting a table of integrals will show it has area one under it, and thus it defines a continuous random variable.

To see what goes astray, let's try to compute the variance of this. This would involve using (A.16) computing

$$\int_{-\infty}^{\infty} \frac{x^2}{1+x^2}\, dx. \tag{A.29}$$

The function on the inside looks like this:

```
# Plot the integrand needed to compute the variance
x = torch.arange(-20, 20, 0.01)
p = x**2 / (1 + x**2)

d2l.plot(x, p, 'x', 'integrand')
```

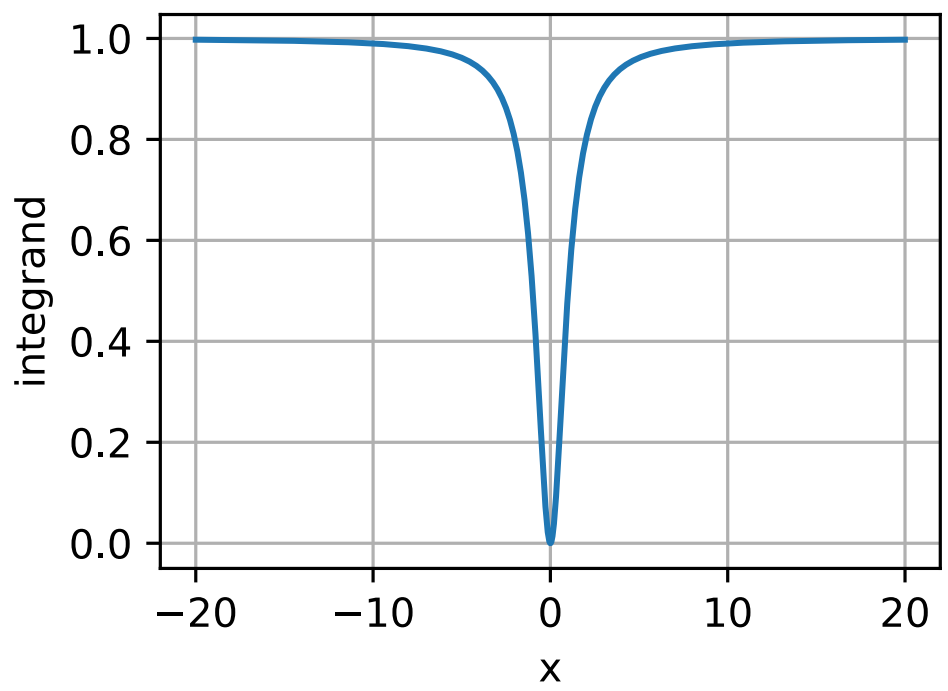

This function clearly has infinite area under it since it is essentially the constant one with a small dip near zero, and indeed we could show that

$$\int_{-\infty}^{\infty} \frac{x^2}{1+x^2}\, dx = \infty. \tag{A.30}$$

This means it does not have a well-defined finite variance.

However, looking deeper shows an even more disturbing result. Let's try to compute the mean using (A.14). Using the change of variables formula, we see

$$\mu_X = \int_{-\infty}^{\infty} \frac{x}{1+x^2}\, dx = \frac{1}{2}\int_1^{\infty} \frac{1}{u}\, du. \tag{A.31}$$

The integral inside is the definition of the logarithm, so this is in essence $\log(\infty) = \infty$, so there is no well-defined average value either!

Machine learning scientists define their models so that we most often do not need to deal with these issues, and will in the vast majority of cases deal with random variables with well-defined means and variances. However, every so often random variables with *heavy tails* (that is those random variables where the probabilities of getting large values are large enough to make things like the mean or variance undefined) are helpful in modeling physical systems, thus it is worth knowing that they exist.

### Joint Density Functions

The above work all assumes we are working with a single real valued random variable. But what if we are dealing with two or more potentially highly correlated random variables? This circumstance is the norm in machine learning: imagine random variables like $R_{i,j}$ which encode the red value of the pixel at the $(i, j)$ coordinate in an image, or $P_t$ which is a random variable given by a stock price at time $t$. Nearby pixels tend to have similar color, and nearby times tend to have similar prices. We cannot treat them as separate random variables, and expect to create a successful model (we will see in Section A.9 a model that under-performs due to such an assumption). We need to develop the mathematical language to handle these correlated continuous random variables.

Thankfully, with the multiple integrals in Section A.5 we can develop such a language. Suppose that we have, for simplicity, two random variables $X, Y$ which can be correlated. Then, similar to the case of a single variable, we can ask the question:

$$P(X \text{ is in an } \epsilon\text{-sized interval around } x \text{ and } Y \text{ is in an } \epsilon\text{-sized interval around } y). \tag{A.32}$$

Similar reasoning to the single variable case shows that this should be approximately

$$P(X \text{ is in an } \epsilon\text{-sized interval around } x \text{ and } Y \text{ is in an } \epsilon\text{-sized interval around } y) \approx \epsilon^{2}p(x, y), \tag{A.33}$$

for some function $p(x, y)$. This is referred to as the joint density of $X$ and $Y$. Similar properties are true for this as we saw in the single variable case. Namely:

- $p(x, y) \geq 0$;
- $\int_{\mathbb{R}^2} p(x, y) \;dx \;dy = 1$;
- $P((X, Y) \in \mathcal{D}) = \int_{\mathcal{D}} p(x, y) \;dx \;dy$.

In this way, we can deal with multiple, potentially correlated random variables. If we wish to work with more than two random variables, we can extend the multivariate density to as many coordinates as desired by considering $p(\mathbf{x}) = p(x_1, \ldots, x_n)$. The same properties of being non-negative, and having total integral of one still hold.

### Marginal Distributions

When dealing with multiple variables, we oftentimes want to be able to ignore the relationships and ask, "how is this one variable distributed?" Such a distribution is called a *marginal distribution*.

To be concrete, let's suppose that we have two random variables $X, Y$ with joint density given by $p_{X,Y}(x, y)$. We will be using the subscript to indicate what random variables the density is for. The question of finding the marginal distribution is taking this function, and using it to find $p_X(x)$.

As with most things, it is best to return to the intuitive picture to figure out what should be true. Recall that the density is the function $p_X$ so that

$$P(X \in [x, x+\epsilon]) \approx \epsilon \cdot p_X(x). \tag{A.34}$$

There is no mention of $Y$, but if all we are given is $p_{X,Y}$, we need to include $Y$ somehow. We can first observe that this is the same as

$$P(X \in [x, x+\epsilon], \text{and } Y \in \mathbb{R}) \approx \epsilon \cdot p_X(x). \tag{A.35}$$

Our density does not directly tell us about what happens in this case, we need to split into small intervals in $y$ as well, so we can write this as

$$\begin{aligned} \epsilon \cdot p_X(x) &\approx \sum_i P(X \in [x, x+\epsilon], \text{and } Y \in [\epsilon \cdot i, \epsilon \cdot (i+1)]) \\ &\approx \sum_i \epsilon^2 p_{X,Y}(x, \epsilon \cdot i). \end{aligned} \tag{A.36}$$

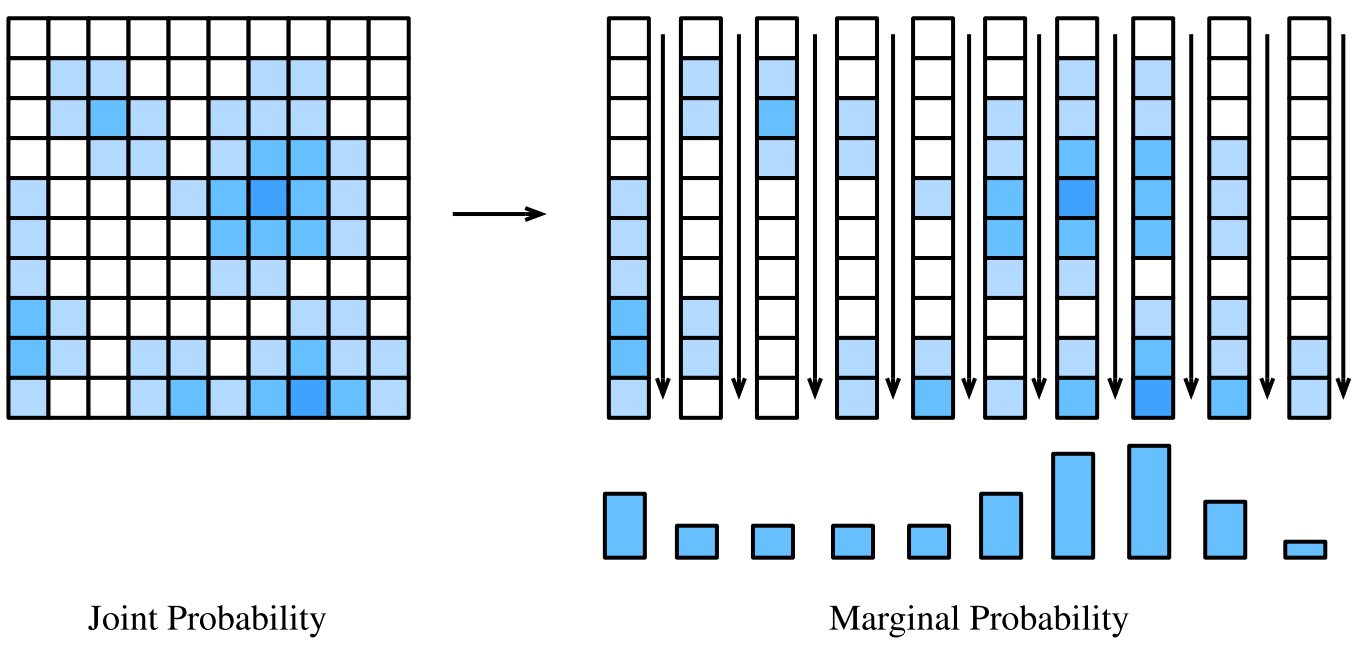

Fig. A.1 By summing along the columns of our array of probabilities, we are able to obtain the marginal distribution for just the random variable represented along the $x$-axis.

This tells us to add up the value of the density along a series of squares in a line as is shown in Fig. A.1. Indeed, after canceling one factor of epsilon from both sides, and recognizing the sum on the right is the integral over $y$, we can conclude that

$$\begin{aligned} p_X(x) &\approx \sum_i \epsilon p_{X,Y}(x, \epsilon \cdot i) \\ &\approx \int_{-\infty}^{\infty} p_{X,Y}(x, y) \; dy. \end{aligned} \tag{A.37}$$

Thus we see

$$p_X(x) = \int_{-\infty}^{\infty} p_{X,Y}(x, y) \; dy. \tag{A.38}$$

This tells us that to get a marginal distribution, we integrate over the variables we do not care about. This process is often referred to as *integrating out* or *marginalized out* the unneeded variables.

## Covariance

When dealing with multiple random variables, there is one additional summary statistic which is helpful to know: the *covariance*. This measures the degree that two random variable fluctuate together.

Suppose that we have two random variables $X$ and $Y$, to begin with, let's suppose they are discrete, taking on values $(x_i, y_j)$ with probability $p_{ij}$. In this case, the covariance is defined as

$$\sigma_{XY} = \mathrm{Cov}(X, Y) = \sum_{i,j} (x_i - \mu_X)(y_j - \mu_Y) p_{ij}. = E[XY] - E[X]E[Y]. \tag{A.39}$$

To think about this intuitively: consider the following pair of random variables. Suppose that $X$ takes the values $1$ and $3$, and $Y$ takes the values $-1$ and $3$. Suppose that we have the following probabilities

$$\begin{aligned}
P(X = 1 \text{ and } Y = -1) &= \frac{p}{2}, \\
P(X = 1 \text{ and } Y = 3) &= \frac{1-p}{2}, \\
P(X = 3 \text{ and } Y = -1) &= \frac{1-p}{2}, \\
P(X = 3 \text{ and } Y = 3) &= \frac{p}{2},
\end{aligned} \tag{A.40}$$

where $p$ is a parameter in $[0, 1]$ we get to pick. Notice that if $p = 1$ then they are both always their minimum or maximum values simultaneously, and if $p = 0$ they are guaranteed to take their flipped values simultaneously (one is large when the other is small and vice versa). If $p = 1/2$, then the four possibilities are all equally likely, and neither should be related. Let's compute the covariance. First, note $\mu_X = 2$ and $\mu_Y = 1$, so we may compute using (A.39):

$$\begin{aligned}
\mathrm{Cov}(X, Y) &= \sum_{i,j} (x_i - \mu_X)(y_j - \mu_Y) p_{ij} \\
&= (1-2)(-1-1)\frac{p}{2} + (1-2)(3-1)\frac{1-p}{2} + (3-2)(-1-1)\frac{1-p}{2} + (3-2)(3-1)\frac{p}{2} \\
&= 4p - 2.
\end{aligned} \tag{A.41}$$

When $p = 1$ (the case where they are both maximally positive or negative at the same time) has a covariance of $2$. When $p = 0$ (the case where they are flipped) the covariance is $-2$. Finally, when $p = 1/2$ (the case where they are unrelated), the covariance is $0$. Thus we see that the covariance measures how these two random variables are related.

A quick note on the covariance is that it only measures these linear relationships. More complex relationships like $X = Y^2$ where $Y$ is randomly chosen from $\{-2, -1, 0, 1, 2\}$ with equal probability can be missed. Indeed a quick computation shows that these random variables have covariance zero, despite one being a deterministic function of the other.

For continuous random variables, much the same story holds. At this point, we are pretty comfortable with doing the transition between discrete and continuous, so we will provide the continuous analogue of (A.39) without any derivation.

$$\sigma_{XY} = \int_{\mathbb{R}^2} (x - \mu_X)(y - \mu_Y)p(x, y) \; dx \; dy. \tag{A.42}$$

For visualization, let's take a look at a collection of random variables with tunable covariance.

```python
# Plot a few random variables adjustable covariance
covs = [-0.9, 0.0, 1.2]
d2l.plt.figure(figsize=(12, 3))
for i in range(3):
    X = torch.randn(500)
    Y = covs[i]*X + torch.randn(500)

    d2l.plt.subplot(1, 4, i+1)
    d2l.plt.scatter(X.numpy(), Y.numpy())
    d2l.plt.xlabel('X')
    d2l.plt.ylabel('Y')
    d2l.plt.title(f'cov = {covs[i]}')
d2l.plt.show()
```

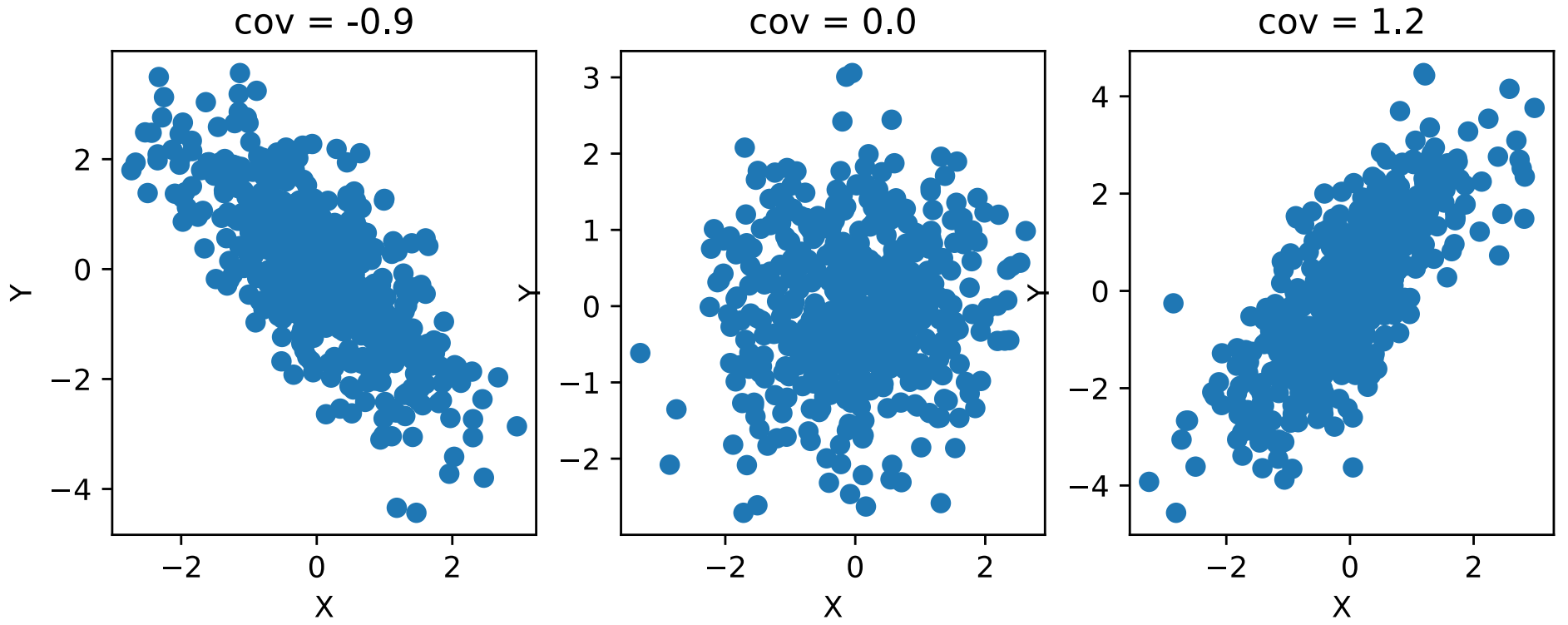

Let's see some properties of covariances:

- For any random variable $X$, $\mathrm{Cov}(X, X) = \mathrm{Var}(X)$.
- For any random variables $X, Y$ and numbers $a$ and $b$, $\mathrm{Cov}(aX + b, Y) = \mathrm{Cov}(X, aY + b) = a\mathrm{Cov}(X, Y)$.
- If $X$ and $Y$ are independent then $\mathrm{Cov}(X, Y) = 0$.

In addition, we can use the covariance to expand a relationship we saw before. Recall that is $X$ and $Y$ are two independent random variables then

$$\mathrm{Var}(X + Y) = \mathrm{Var}(X) + \mathrm{Var}(Y). \tag{A.43}$$

With knowledge of covariances, we can expand this relationship. Indeed, some algebra can show that in general,

$$\mathrm{Var}(X+Y) = \mathrm{Var}(X) + \mathrm{Var}(Y) + 2\mathrm{Cov}(X,Y). \tag{A.44}$$

This allows us to generalize the variance summation rule for correlated random variables.

## Correlation

As we did in the case of means and variances, let's now consider units. If $X$ is measured in one unit (say inches), and $Y$ is measured in another (say dollars), the covariance is measured in the product of these two units inches $\times$ dollars. These units can be hard to interpret. What we will often want in this case is a unit-less measurement of relatedness. Indeed, often we do not care about exact quantitative correlation, but rather ask if the correlation is in the same direction, and how strong the relationship is.

To see what makes sense, let's perform a thought experiment. Suppose that we convert our random variables in inches and dollars to be in inches and cents. In this case the random variable $Y$ is multiplied by $100$. If we work through the definition, this means that $\mathrm{Cov}(X,Y)$ will be multiplied by $100$. Thus we see that in this case a change of units change the covariance by a factor of $100$. Thus, to find our unit-invariant measure of correlation, we will need to divide by something else that also gets scaled by $100$. Indeed we have a clear candidate, the standard deviation! Indeed if we define the *correlation coefficient* to be

$$\rho(X,Y) = \frac{\mathrm{Cov}(X,Y)}{\sigma_X \sigma_Y}, \tag{A.45}$$

we see that this is a unit-less value. A little mathematics can show that this number is between $-1$ and $1$ with $1$ meaning maximally positively correlated, whereas $-1$ means maximally negatively correlated.

Returning to our explicit discrete example above, we can see that $\sigma_X = 1$ and $\sigma_Y = 2$, so we can compute the correlation between the two random variables using (A.45) to see that

$$\rho(X,Y) = \frac{4p-2}{1\cdot 2} = 2p-1. \tag{A.46}$$

This now ranges between $-1$ and $1$ with the expected behavior of $1$ meaning most correlated, and $-1$ meaning minimally correlated.

As another example, consider $X$ as any random variable, and $Y = aX + b$ as any linear deterministic function of $X$. Then, one can compute that

$$\sigma_Y = \sigma_{aX+b} = |a|\sigma_X, \tag{A.47}$$

$$\mathrm{Cov}(X,Y) = \mathrm{Cov}(X, aX+b) = a\mathrm{Cov}(X,X) = a\mathrm{Var}(X), \tag{A.48}$$

and thus by (A.45) that

$$\rho(X,Y) = \frac{a\mathrm{Var}(X)}{|a|\sigma_X^2} = \frac{a}{|a|} = \mathrm{sign}(a). \tag{A.49}$$

Thus we see that the correlation is $+1$ for any $a > 0$, and $-1$ for any $a < 0$ illustrating that correlation measures the degree and directionality the two random variables are related, not the scale that the variation takes.

Let's again plot a collection of random variables with tunable correlation.

```
# Plot a few random variables adjustable correlations
cors = [-0.9, 0.0, 1.0]
d2l.plt.figure(figsize=(12, 3))
for i in range(3):
    X = torch.randn(500)
    Y = cors[i] * X + torch.sqrt(torch.tensor(1) -
                                 cors[i]**2) * torch.randn(500)

    d2l.plt.subplot(1, 4, i + 1)
    d2l.plt.scatter(X.numpy(), Y.numpy())
    d2l.plt.xlabel('X')
    d2l.plt.ylabel('Y')
    d2l.plt.title(f'cor = {cors[i]}')
d2l.plt.show()
```

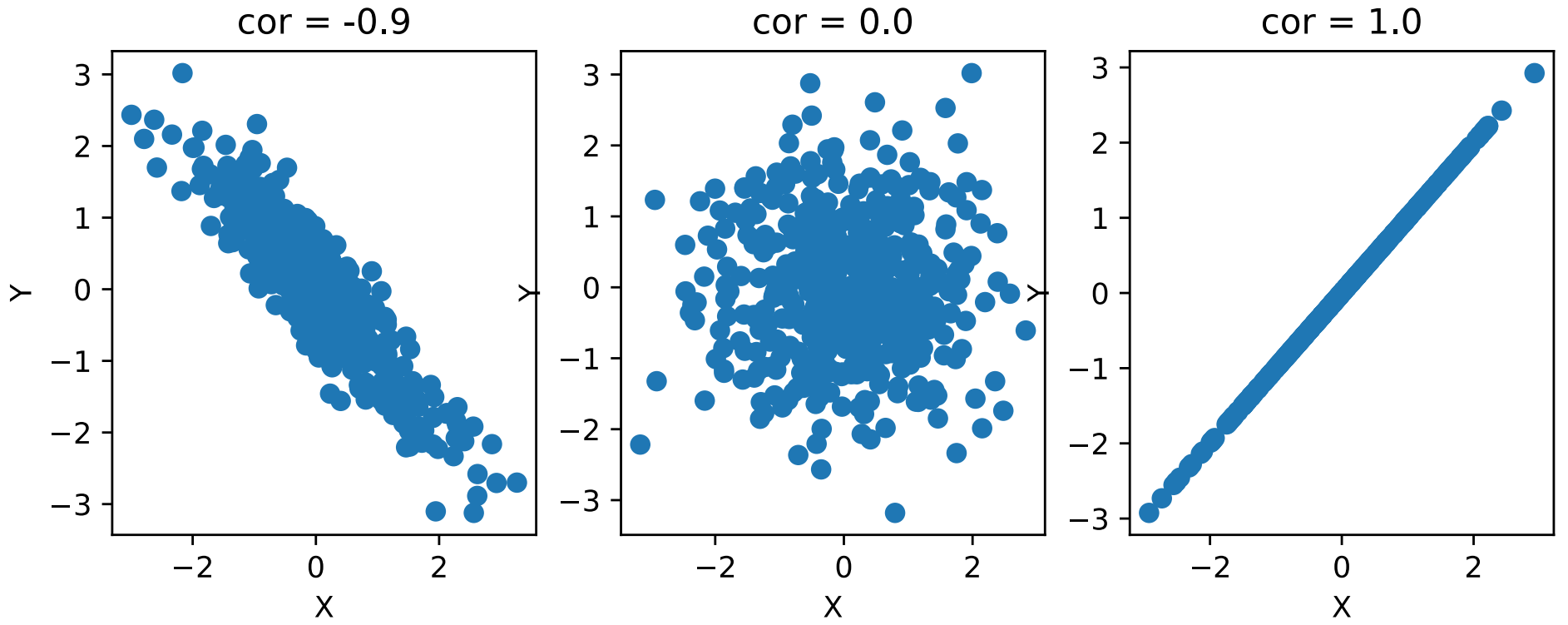

Let's list a few properties of the correlation below.

- For any random variable $X$, $\rho(X, X) = 1$.
- For any random variables $X, Y$ and numbers $a$ and $b$, $\rho(aX + b, Y) = \rho(X, aY + b) = \rho(X, Y)$.
- If $X$ and $Y$ are independent with non-zero variance then $\rho(X, Y) = 0$.

As a final note, you may feel like some of these formulae are familiar. Indeed, if we expand everything out assuming that $\mu_X = \mu_Y = 0$, we see that this is

$$\rho(X, Y) = \frac{\sum_{i,j} x_i y_i p_{ij}}{\sqrt{\sum_{i,j} x_i^2 p_{ij}} \sqrt{\sum_{i,j} y_j^2 p_{ij}}}. \tag{A.50}$$

This looks like a sum of a product of terms divided by the square root of sums of terms. This is exactly the formula for the cosine of the angle between two vectors $\mathbf{v}, \mathbf{w}$ with the different coordinates weighted by $p_{ij}$:

$$\cos(\theta) = \frac{\mathbf{v} \cdot \mathbf{w}}{\|\mathbf{v}\|\|\mathbf{w}\|} = \frac{\sum_i v_i w_i}{\sqrt{\sum_i v_i^2}\sqrt{\sum_i w_i^2}}. \tag{A.51}$$

Indeed if we think of norms as being related to standard deviations, and correlations as being cosines of angles, much of the intuition we have from geometry can be applied to thinking about random variables.

## A.6.2 Summary

- Continuous random variables are random variables that can take on a continuum of values. They have some technical difficulties that make them more challenging to work with compared to discrete random variables.
- The probability density function allows us to work with continuous random variables by giving a function where the area under the curve on some interval gives the probability of finding a sample point in that interval.
- The cumulative distribution function is the probability of observing the random variable to be less than a given threshold. It can provide a useful alternate viewpoint which unifies discrete and continuous variables.
- The mean is the average value of a random variable.
- The variance is the expected square of the difference between the random variable and its mean.
- The standard deviation is the square root of the variance. It can be thought of as measuring the range of values the random variable may take.
- Chebyshev's inequality allows us to make this intuition rigorous by giving an explicit interval that contains the random variable most of the time.
- Joint densities allow us to work with correlated random variables. We may marginalize joint densities by integrating over unwanted random variables to get the distribution of the desired random variable.
- The covariance and correlation coefficient provide a way to measure any linear relationship between two correlated random variables.

## A.6.3 Exercises

1. Suppose that we have the random variable with density given by $p(x) = \frac{1}{x^2}$ for $x \geq 1$ and $p(x) = 0$ otherwise. What is $P(X > 2)$?

2. The Laplace distribution is a random variable whose density is given by $p(x = \frac{1}{2}e^{-|x|}$. What is the mean and the standard deviation of this function? As a hint, $\int_0^\infty xe^{-x}\, dx = 1$ and $\int_0^\infty x^2 e^{-x}\, dx = 2$.
3. I walk up to you on the street and say "I have a random variable with mean $1$, standard deviation $2$, and I observed $25\%$ of my samples taking a value larger than $9$." Do you believe me? Why or why not?
4. Suppose that you have two random variables $X, Y$, with joint density given by $p_{XY}(x, y) = 4xy$ for $x, y \in [0, 1]$ and $p_{XY}(x, y) = 0$ otherwise. What is the covariance of $X$ and $Y$?

Discussions[283].

# A.7 Maximum Likelihood

One of the most commonly encountered way of thinking in machine learning is the maximum likelihood point of view. This is the concept that when working with a probabilistic model with unknown parameters, the parameters which make the data have the highest probability are the most likely ones.

## A.7.1 The Maximum Likelihood Principle

This has a Bayesian interpretation which can be helpful to think about. Suppose that we have a model with parameters $\boldsymbol{\theta}$ and a collection of data examples $X$. For concreteness, we can imagine that $\boldsymbol{\theta}$ is a single value representing the probability that a coin comes up heads when flipped, and $X$ is a sequence of independent coin flips. We will look at this example in depth later.

If we want to find the most likely value for the parameters of our model, that means we want to find

$$\mathrm{argmax}\, P(\boldsymbol{\theta} \mid X). \tag{A.1}$$

By Bayes' rule, this is the same thing as

$$\mathrm{argmax}\, \frac{P(X \mid \boldsymbol{\theta})P(\boldsymbol{\theta})}{P(X)}. \tag{A.2}$$

The expression $P(X)$, a parameter agnostic probability of generating the data, does not depend on $\boldsymbol{\theta}$ at all, and so can be dropped without changing the best choice of $\boldsymbol{\theta}$. Similarly, we may now posit that we have no prior assumption on which set of parameters are better than any others, so we may declare that $P(\boldsymbol{\theta})$ does not depend on theta either! This, for instance, makes sense in our coin flipping example where the probability it comes up heads could be any value in $[0, 1]$ without any prior belief it is fair or not (often referred to as an *uninformative prior*). Thus we see that our application of Bayes' rule shows that our best choice of $\boldsymbol{\theta}$ is the maximum likelihood estimate for $\boldsymbol{\theta}$:

$$\hat{\boldsymbol{\theta}} = \underset{\boldsymbol{\theta}}{\operatorname{argmax}} \, P(X \mid \boldsymbol{\theta}). \tag{A.3}$$

As a matter of common terminology, the probability of the data given the parameters ($P(X \mid \boldsymbol{\theta})$) is referred to as the *likelihood*.

## A Concrete Example

Let's see how this works in a concrete example. Suppose that we have a single parameter $\theta$ representing the probability that a coin flip is heads. Then the probability of getting a tails is $1 - \theta$, and so if our observed data $X$ is a sequence with $n_H$ heads and $n_T$ tails, we can use the fact that independent probabilities multiply to see that

$$P(X \mid \theta) = \theta^{n_H}(1 - \theta)^{n_T}. \tag{A.4}$$

If we flip 13 coins and get the sequence "HHHTHTTHHHHHT", which has $n_H = 9$ and $n_T = 4$, we see that this is

$$P(X \mid \theta) = \theta^9(1 - \theta)^4. \tag{A.5}$$

One nice thing about this example will be that we know the answer going in. Indeed, if we said verbally, "I flipped 13 coins, and 9 came up heads, what is our best guess for the probability that the coin comes us heads?, " everyone would correctly guess $9/13$. What this maximum likelihood method will give us is a way to get that number from first principals in a way that will generalize to vastly more complex situations.

For our example, the plot of $P(X \mid \theta)$ is as follows:

```
%matplotlib inline
import torch
from d2l import torch as d2l

theta = torch.arange(0, 1, 0.001)
p = theta**9 * (1 - theta)**4.

d2l.plot(theta, p, 'theta', 'likelihood')
```

This has its maximum value somewhere near our expected $9/13 \approx 0.7\ldots$. To see if it is exactly there, we can turn to calculus. Notice that at the maximum, the gradient of the function is flat. Thus, we could find the maximum likelihood estimate (A.1) by finding the values of $\theta$ where the derivative is zero, and finding the one that gives the highest probability.

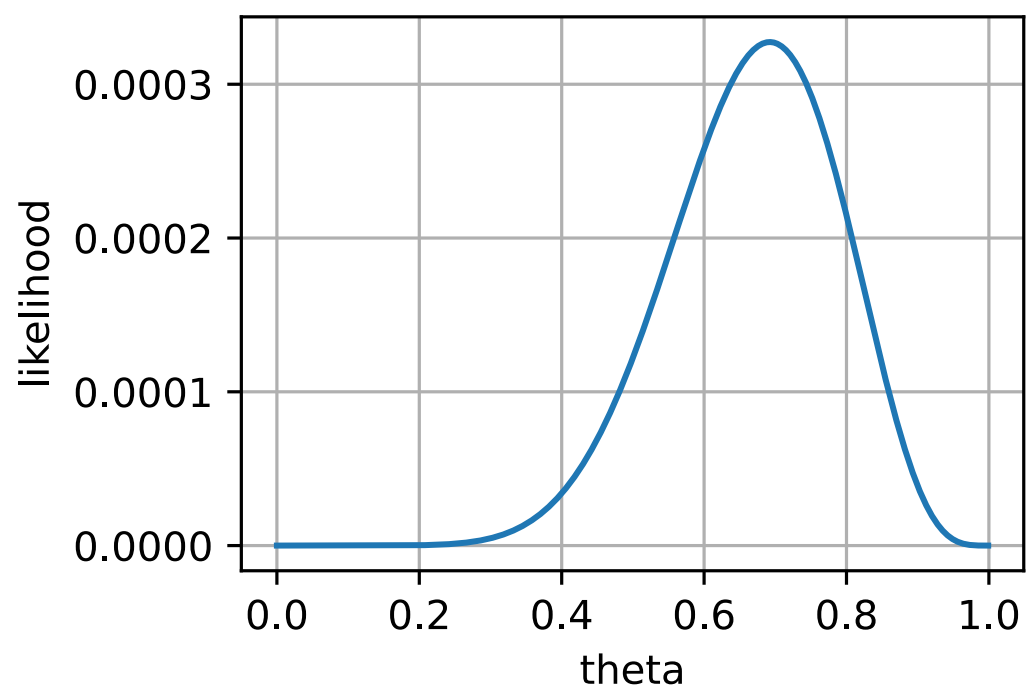

We compute:

$$\begin{aligned} 0 &= \frac{d}{d\theta} P(X \mid \theta) \\ &= \frac{d}{d\theta} \theta^9 (1-\theta)^4 \\ &= 9\theta^8 (1-\theta)^4 - 4\theta^9 (1-\theta)^3 \\ &= \theta^8 (1-\theta)^3 (9 - 13\theta). \end{aligned} \tag{A.6}$$

This has three solutions: $0$, $1$ and $9/13$. The first two are clearly minima, not maxima as they assign probability $0$ to our sequence. The final value does *not* assign zero probability to our sequence, and thus must be the maximum likelihood estimate $\hat{\theta} = 9/13$.

## A.7.2 Numerical Optimization and the Negative Log-Likelihood

The previous example is nice, but what if we have billions of parameters and data examples?

First, notice that if we make the assumption that all the data examples are independent, we can no longer practically consider the likelihood itself as it is a product of many probabilities. Indeed, each probability is in $[0, 1]$, say typically of value about $1/2$, and the product of $(1/2)^{1000000000}$ is far below machine precision. We cannot work with that directly.

However, recall that the logarithm turns products to sums, in which case

$$\log((1/2)^{1000000000}) = 1000000000 \cdot \log(1/2) \approx -301029995.6\ldots \tag{A.7}$$

This number fits perfectly within even a single precision 32-bit float. Thus, we should consider the *log-likelihood*, which is

$$\log(P(X \mid \boldsymbol{\theta})). \tag{A.8}$$

Since the function $x \mapsto \log(x)$ is increasing, maximizing the likelihood is the same thing as maximizing the log-likelihood. Indeed in Section A.9 we will see this reasoning applied when working with the specific example of the naive Bayes classifier.

We often work with loss functions, where we wish to minimize the loss. We may turn maximum likelihood into the minimization of a loss by taking $-\log(P(X \mid \boldsymbol{\theta}))$, which is the *negative log-likelihood*.

To illustrate this, consider the coin flipping problem from before, and pretend that we do not know the closed form solution. We may compute that

$$-\log(P(X \mid \boldsymbol{\theta})) = -\log(\theta^{n_H}(1-\theta)^{n_T}) = -(n_H \log(\theta) + n_T \log(1-\theta)). \tag{A.9}$$

This can be written into code, and freely optimized even for billions of coin flips.

```
# Set up our data
n_H = 8675309
n_T = 256245

# Initialize our paramteres
theta = torch.tensor(0.5, requires_grad=True)

# Perform gradient descent
lr = 1e-9
for iter in range(100):
    loss = -(n_H * torch.log(theta) + n_T * torch.log(1 - theta))
    loss.backward()
    with torch.no_grad():
        theta -= lr * theta.grad
    theta.grad.zero_()

# Check output
theta, n_H / (n_H + n_T)
```

```
(tensor(0.9713, requires_grad=True), 0.9713101437890875)
```

Numerical convenience is not the only reason why people like to use negative log-likelihoods. There are several other reasons why it is preferable.

The second reason we consider the log-likelihood is the simplified application of calculus rules. As discussed above, due to independence assumptions, most probabilities we encounter in machine learning are products of individual probabilities.

$$P(X \mid \boldsymbol{\theta}) = p(x_1 \mid \boldsymbol{\theta}) \cdot p(x_2 \mid \boldsymbol{\theta}) \cdots p(x_n \mid \boldsymbol{\theta}). \tag{A.10}$$

This means that if we directly apply the product rule to compute a derivative we get

$$\begin{aligned} \frac{\partial}{\partial \boldsymbol{\theta}} P(X \mid \boldsymbol{\theta}) &= \left(\frac{\partial}{\partial \boldsymbol{\theta}} P(x_1 \mid \boldsymbol{\theta})\right) \cdot P(x_2 \mid \boldsymbol{\theta}) \cdots P(x_n \mid \boldsymbol{\theta}) \\ &\quad + P(x_1 \mid \boldsymbol{\theta}) \cdot \left(\frac{\partial}{\partial \boldsymbol{\theta}} P(x_2 \mid \boldsymbol{\theta})\right) \cdots P(x_n \mid \boldsymbol{\theta}) \\ &\quad \vdots \\ &\quad + P(x_1 \mid \boldsymbol{\theta}) \cdot P(x_2 \mid \boldsymbol{\theta}) \cdots \left(\frac{\partial}{\partial \boldsymbol{\theta}} P(x_n \mid \boldsymbol{\theta})\right). \end{aligned} \tag{A.11}$$

This requires $n(n-1)$ multiplications, along with $(n-1)$ additions, so it is proportional to quadratic time in the inputs! Sufficient cleverness in grouping terms will reduce this to linear time, but it requires some thought. For the negative log-likelihood we have instead

$$-\log\left(P(X \mid \boldsymbol{\theta})\right) = -\log(P(x_1 \mid \boldsymbol{\theta})) - \log(P(x_2 \mid \boldsymbol{\theta})) \cdots - \log(P(x_n \mid \boldsymbol{\theta})), \tag{A.12}$$

which then gives

$$-\frac{\partial}{\partial \boldsymbol{\theta}} \log\left(P(X \mid \boldsymbol{\theta})\right) = \frac{1}{P(x_1 \mid \boldsymbol{\theta})}\left(\frac{\partial}{\partial \boldsymbol{\theta}} P(x_1 \mid \boldsymbol{\theta})\right) + \cdots + \frac{1}{P(x_n \mid \boldsymbol{\theta})}\left(\frac{\partial}{\partial \boldsymbol{\theta}} P(x_n \mid \boldsymbol{\theta})\right). \tag{A.13}$$

This requires only $n$ divides and $n-1$ sums, and thus is linear time in the inputs.

The third and final reason to consider the negative log-likelihood is the relationship to information theory, which we will discuss in detail in Section A.11. This is a rigorous mathematical theory which gives a way to measure the degree of information or randomness in a random variable. The key object of study in that field is the entropy which is

$$H(p) = -\sum_i p_i \log_2(p_i), \tag{A.14}$$

which measures the randomness of a source. Notice that this is nothing more than the average $-\log$ probability, and thus if we take our negative log-likelihood and divide by the number of data examples, we get a relative of entropy known as cross-entropy. This theoretical interpretation alone would be sufficiently compelling to motivate reporting the average negative log-likelihood over the dataset as a way of measuring model performance.

### A.7.3 Maximum Likelihood for Continuous Variables

Everything that we have done so far assumes we are working with discrete random variables, but what if we want to work with continuous ones?

The short summary is that nothing at all changes, except we replace all the instances of the probability with the probability density. Recalling that we write densities with lower case $p$, this means that for example we now say

$$-\log\left(p(X \mid \boldsymbol{\theta})\right) = -\log(p(x_1 \mid \boldsymbol{\theta})) - \log(p(x_2 \mid \boldsymbol{\theta})) \cdots - \log(p(x_n \mid \boldsymbol{\theta})) = -\sum_i \log(p(x_i \mid \theta)). \tag{A.15}$$

The question becomes, "Why is this OK?" After all, the reason we introduced densities was because probabilities of getting specific outcomes themselves was zero, and thus is not the probability of generating our data for any set of parameters zero?

Indeed, this is the case, and understanding why we can shift to densities is an exercise in tracing what happens to the epsilons.

Let's first re-define our goal. Suppose that for continuous random variables we no longer want to compute the probability of getting exactly the right value, but instead matching to within some range $\epsilon$. For simplicity, we assume our data is repeated observations $x_1, \ldots, x_N$ of identically distributed random variables $X_1, \ldots, X_N$. As we have seen previously, this can be written as

$$\begin{aligned} &P(X_1 \in [x_1, x_1 + \epsilon], X_2 \in [x_2, x_2 + \epsilon], \ldots, X_N \in [x_N, x_N + \epsilon] \mid \boldsymbol{\theta}) \\ \approx &\epsilon^N p(x_1 \mid \boldsymbol{\theta}) \cdot p(x_2 \mid \boldsymbol{\theta}) \cdots p(x_n \mid \boldsymbol{\theta}). \end{aligned} \tag{A.16}$$

Thus, if we take negative logarithms of this we obtain

$$\begin{aligned} &-\log(P(X_1 \in [x_1, x_1 + \epsilon], X_2 \in [x_2, x_2 + \epsilon], \ldots, X_N \in [x_N, x_N + \epsilon] \mid \boldsymbol{\theta})) \\ \approx &-N \log(\epsilon) - \sum_{i} \log(p(x_i \mid \boldsymbol{\theta})). \end{aligned} \tag{A.17}$$

If we examine this expression, the only place that the $\epsilon$ occurs is in the additive constant $-N \log(\epsilon)$. This does not depend on the parameters $\boldsymbol{\theta}$ at all, so the optimal choice of $\boldsymbol{\theta}$ does not depend on our choice of $\epsilon$! If we demand four digits or four-hundred, the best choice of $\boldsymbol{\theta}$ remains the same, thus we may freely drop the epsilon to see that what we want to optimize is

$$-\sum_{i} \log(p(x_i \mid \boldsymbol{\theta})). \tag{A.18}$$

Thus, we see that the maximum likelihood point of view can operate with continuous random variables as easily as with discrete ones by replacing the probabilities with probability densities.

### A.7.4 Summary

- The maximum likelihood principle tells us that the best fit model for a given dataset is the one that generates the data with the highest probability.
- Often people work with the negative log-likelihood instead for a variety of reasons: numerical stability, conversion of products to sums (and the resulting simplification of gradient computations), and theoretical ties to information theory.
- While simplest to motivate in the discrete setting, it may be freely generalized to the continuous setting as well by maximizing the probability density assigned to the datapoints.

### A.7.5 Exercises

1. Suppose that you know that a non-negative random variable has density $\alpha e^{-\alpha x}$ for some value $\alpha > 0$. You obtain a single observation from the random variable which is the number 3. What is the maximum likelihood estimate for $\alpha$?
2. Suppose that you have a dataset of samples $\{x_i\}_{i=1}^N$ drawn from a Gaussian with unknown mean, but variance 1. What is the maximum likelihood estimate for the mean?

Discussions[284].

# A.8 Distributions

Now that we have learned how to work with probability in both the discrete and the continuous setting, let's get to know some of the common distributions encountered. Depending on the area of machine learning, we may need to be familiar with vastly more of these, or for some areas of deep learning potentially none at all. This is, however, a good basic list to be familiar with. Let's first import some common libraries.

```
%matplotlib inline
from math import erf, factorial
import torch
from IPython import display
from d2l import torch as d2l

torch.pi = torch.acos(torch.zeros(1)) * 2  # Define pi in torch
```

## A.8.1 Bernoulli

This is the simplest random variable usually encountered. This random variable encodes a coin flip which comes up 1 with probability $p$ and 0 with probability $1 - p$. If we have a random variable $X$ with this distribution, we will write

$$X \sim \mathrm{Bernoulli}(p). \tag{A.1}$$

The cumulative distribution function is

$$F(x) = \begin{cases} 0 & x < 0, \\ 1 - p & 0 \leq x < 1, \\ 1 & x >= 1 . \end{cases} \tag{A.2}$$

The probability mass function is plotted below.

```
p = 0.3

d2l.set_figsize()
d2l.plt.stem([0, 1], [1 - p, p], use_line_collection=True)
d2l.plt.xlabel('x')
d2l.plt.ylabel('p.m.f.')
d2l.plt.show()
```

Now, let's plot the cumulative distribution function (A.2).

```
x = torch.arange(-1, 2, 0.01)

def F(x):
```

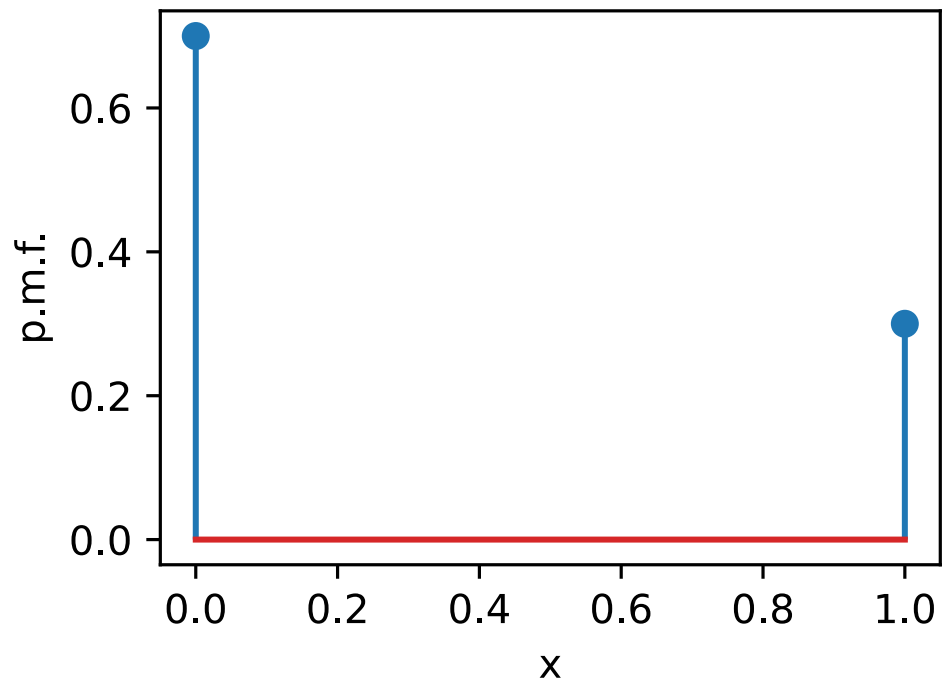

```
    return 0 if x < 0 else 1 if x > 1 else 1 - p

d2l.plot(x, torch.tensor([F(y) for y in x]), 'x', 'c.d.f.')
```

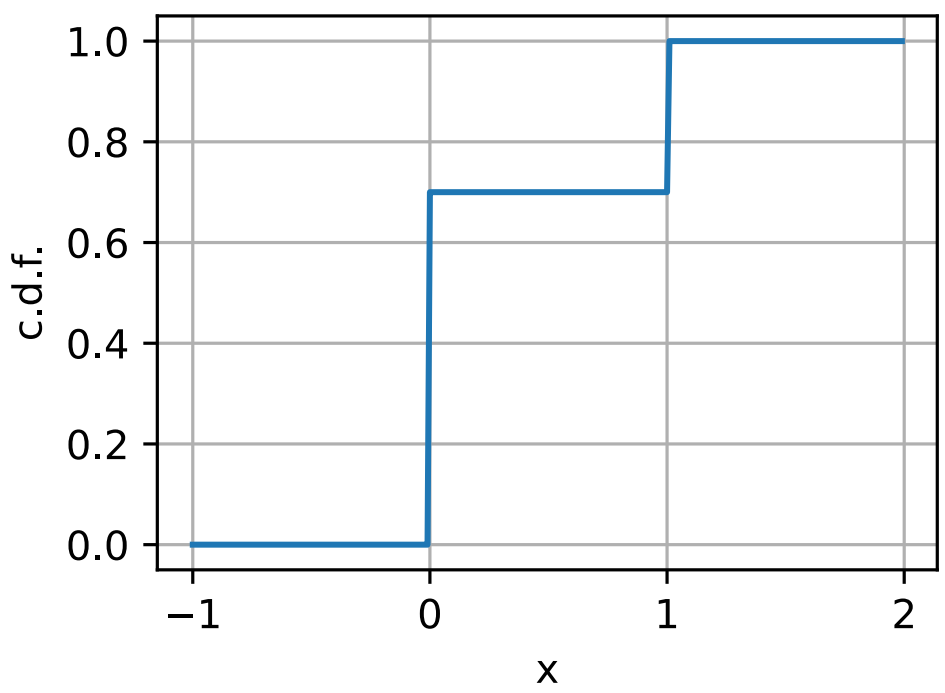

If $X \sim \mathrm{Bernoulli}(p)$, then:

- $\mu_X = p$,
- $\sigma_X^2 = p(1-p)$.

We can sample an array of arbitrary shape from a Bernoulli random variable as follows.

```
1*(torch.rand(10, 10) < p)
```

```
tensor([[0, 0, 0, 0, 0, 1, 0, 1, 1, 0],
        [1, 0, 1, 1, 0, 0, 0, 0, 1, 1],
        [1, 0, 0, 1, 0, 0, 0, 0, 0, 1],
        [1, 1, 0, 0, 0, 0, 0, 1, 0, 0],
        [1, 0, 1, 1, 1, 0, 0, 0, 1, 0],
        [1, 0, 0, 1, 0, 0, 0, 0, 0, 0],
        [1, 0, 1, 1, 0, 0, 1, 0, 0, 0],
```

```
        [1, 0, 0, 0, 0, 0, 0, 0, 0, 0],
        [1, 1, 1, 0, 0, 1, 0, 0, 0, 1],
        [0, 0, 0, 0, 1, 0, 0, 0, 0, 0]])
```

## A.8.2 Discrete Uniform

The next commonly encountered random variable is a discrete uniform. For our discussion here, we will assume that it is supported on the integers $\{1, 2, \ldots, n\}$, however any other set of values can be freely chosen. The meaning of the word *uniform* in this context is that every possible value is equally likely. The probability for each value $i \in \{1, 2, 3, \ldots, n\}$ is $p_i = \frac{1}{n}$. We will denote a random variable $X$ with this distribution as

$$X \sim U(n). \tag{A.3}$$

The cumulative distribution function is

$$F(x) = \begin{cases} 0 & x < 1, \\ \frac{k}{n} & k \leq x < k+1 \text{ with } 1 \leq k < n, \\ 1 & x >= n. \end{cases} \tag{A.4}$$

Let's first plot the probability mass function.

```
n = 5

d2l.plt.stem([i+1 for i in range(n)], n*[1 / n], use_line_collection=True)
d2l.plt.xlabel('x')
d2l.plt.ylabel('p.m.f.')
d2l.plt.show()
```

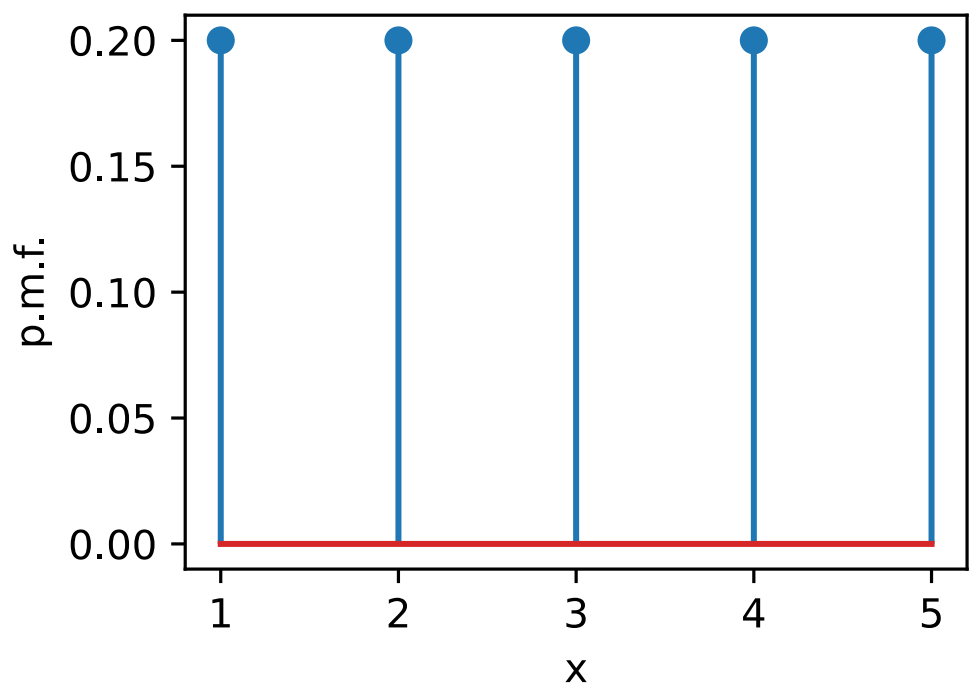

Now, let's plot the cumulative distribution function (A.4).

```
x = torch.arange(-1, 6, 0.01)

def F(x):
    return 0 if x < 1 else 1 if x > n else torch.floor(x) / n

d2l.plot(x, torch.tensor([F(y) for y in x]), 'x', 'c.d.f.')
```

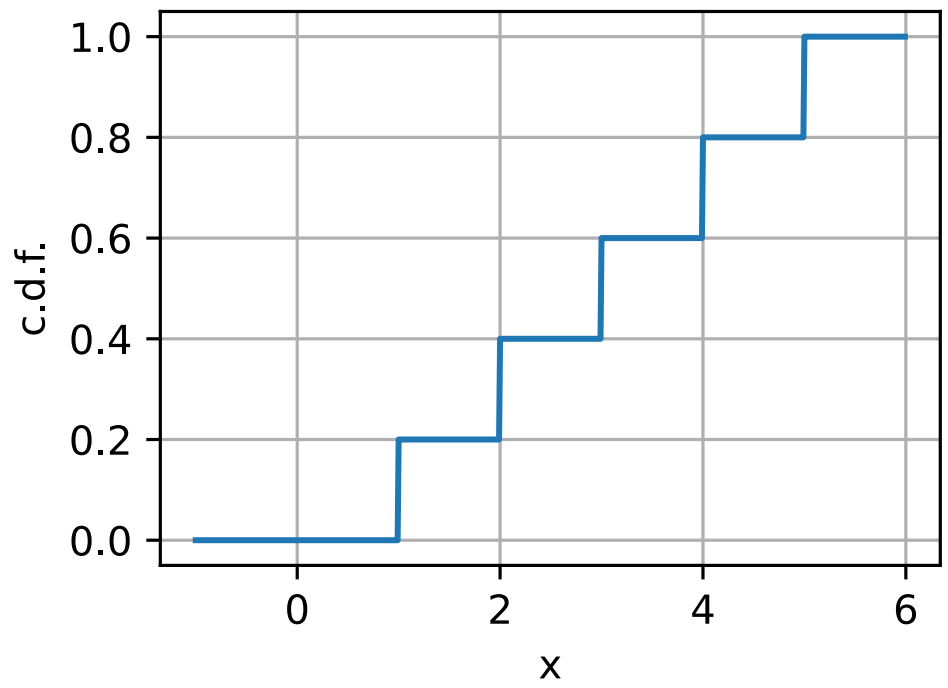

If $X \sim U(n)$, then:

- $\mu_X = \frac{1+n}{2}$,
- $\sigma_X^2 = \frac{n^2-1}{12}$.

We can sample an array of arbitrary shape from a discrete uniform random variable as follows.

```
torch.randint(1, n, size=(10, 10))
```

```
tensor([[1, 3, 4, 1, 1, 1, 2, 2, 1, 3],
        [1, 1, 1, 2, 2, 1, 1, 2, 2, 3],
        [2, 3, 2, 2, 4, 3, 2, 2, 3, 2],
        [1, 2, 1, 4, 3, 2, 3, 2, 4, 3],
        [1, 2, 2, 3, 2, 1, 2, 2, 3, 3],
        [2, 4, 1, 1, 2, 1, 1, 2, 3, 4],
        [3, 1, 3, 1, 1, 3, 4, 4, 1, 4],
        [1, 1, 3, 1, 1, 4, 3, 3, 4, 1],
        [1, 2, 3, 2, 4, 4, 3, 4, 3, 3],
        [2, 1, 1, 2, 1, 3, 2, 3, 1, 4]])
```

### A.8.3 Continuous Uniform

Next, let's discuss the continuous uniform distribution. The idea behind this random variable is that if we increase the $n$ in the discrete uniform distribution, and then scale it to fit within the interval $[a, b]$, we will approach a continuous random variable that just picks an arbitrary value in $[a, b]$ all with equal probability. We will denote this distribution as

$$X \sim U(a, b). \tag{A.5}$$

The probability density function is

$$p(x) = \begin{cases} \frac{1}{b-a} & x \in [a, b], \\ 0 & x \notin [a, b]. \end{cases} \tag{A.6}$$

The cumulative distribution function is

$$F(x) = \begin{cases} 0 & x < a, \\ \frac{x-a}{b-a} & x \in [a, b], \\ 1 & x >= b. \end{cases} \tag{A.7}$$

Let's first plot the probability density function (A.6).

```
a, b = 1, 3

x = torch.arange(0, 4, 0.01)
p = (x > a).type(torch.float32)*(x < b).type(torch.float32)/(b-a)
d2l.plot(x, p, 'x', 'p.d.f.')
```

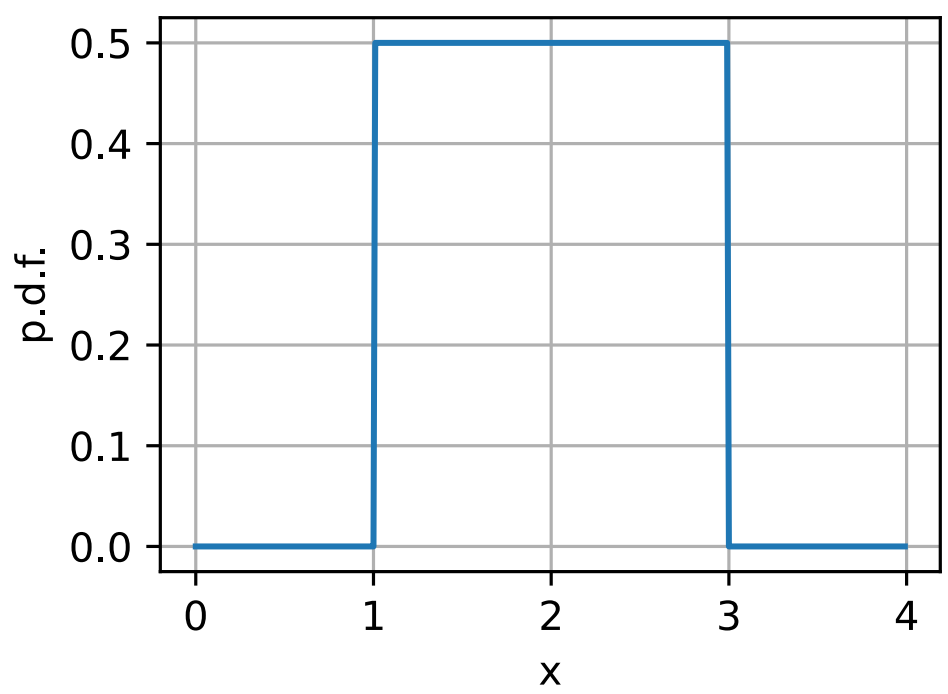

Now, let's plot the cumulative distribution function (A.7).

```
def F(x):
    return 0 if x < a else 1 if x > b else (x - a) / (b - a)

d2l.plot(x, torch.tensor([F(y) for y in x]), 'x', 'c.d.f.')
```

If $X \sim U(a, b)$, then:

- $\mu_X = \frac{a+b}{2}$,
- $\sigma_X^2 = \frac{(b-a)^2}{12}$.

We can sample an array of arbitrary shape from a uniform random variable as follows. Note that it by default samples from a $U(0, 1)$, so if we want a different range we need to scale it.

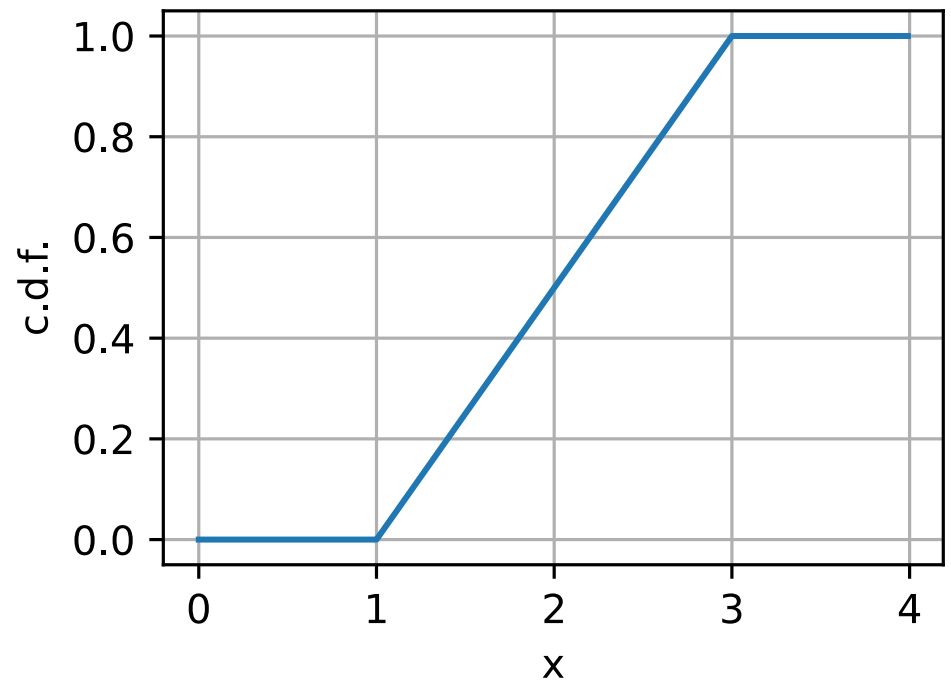

```
(b - a) * torch.rand(10, 10) + a
```

```
tensor([[1.7669, 1.1279, 1.4392, 2.4289, 1.6923, 1.3447, 2.7770, 1.3319, 2.
↪2326,
         1.3150],
        [2.7011, 2.8355, 1.5303, 2.0331, 2.8806, 1.6149, 2.5169, 2.9279, 1.
↪3378,
         2.5879],
        [1.9821, 1.5423, 2.6820, 1.2674, 1.7578, 1.6478, 2.7660, 1.4664, 1.
↪1517,
         1.6246],
        [2.1729, 1.9362, 1.2715, 2.9132, 1.3799, 1.8364, 1.5225, 2.3158, 1.
↪9384,
         2.3300],
        [2.3944, 1.8187, 2.3988, 1.6223, 1.2805, 2.8716, 2.8287, 1.2452, 1.
↪8449,
         2.4097],
        [1.3478, 2.8731, 2.4662, 2.0276, 1.1287, 2.8155, 1.5838, 1.2206, 2.
↪2059,
         2.8454],
        [2.4784, 2.5221, 1.1688, 2.7545, 2.6526, 2.9883, 2.1063, 1.6344, 2.
↪4662,
         1.8148],
        [1.6195, 2.8807, 1.6211, 1.2088, 1.9361, 2.0862, 2.3551, 1.7576, 1.
↪6081,
         1.5808],
        [2.6168, 2.6342, 2.3619, 2.5629, 2.0102, 2.0918, 1.6956, 1.6710, 1.
↪4272,
         1.7900],
        [2.2539, 2.6632, 2.0519, 1.5298, 2.2536, 2.0400, 1.8634, 1.5086, 2.
↪0670,
         1.6122]])
```

### A.8.4 Binomial

Let's make things a little more complex and examine the *binomial* random variable. This random variable originates from performing a sequence of $n$ independent experiments, each of which has probability $p$ of succeeding, and asking how many successes we expect to see.

Let's express this mathematically. Each experiment is an independent random variable $X_i$ where we will use 1 to encode success, and 0 to encode failure. Since each is an independent coin flip which is successful with probability $p$, we can say that $X_i \sim \textrm{Bernoulli}(p)$. Then, the binomial random variable is

$$X = \sum_{i=1}^n X_i. \tag{A.8}$$

In this case, we will write

$$X \sim \textrm{Binomial}(n, p). \tag{A.9}$$

To get the cumulative distribution function, we need to notice that getting exactly $k$ successes can occur in $\binom{n}{k} = \frac{n!}{k!(n-k)!}$ ways each of which has a probability of $p^k(1-p)^{n-k}$ of occurring. Thus the cumulative distribution function is

$$F(x) = \begin{cases} 0 & x < 0, \\ \sum_{m \le k} \binom{n}{m} p^m(1-p)^{n-m} & k \le x < k+1 \textrm{ with } 0 \le k < n, \\ 1 & x >= n. \end{cases} \tag{A.10}$$

Let's first plot the probability mass function.

```
n, p = 10, 0.2

# Compute binomial coefficient
def binom(n, k):
    comb = 1
    for i in range(min(k, n - k)):
        comb = comb * (n - i) // (i + 1)
    return comb

pmf = torch.tensor([p**i * (1-p)**(n - i) * binom(n, i) for i in range(n + 1)])

d2l.plt.stem([i for i in range(n + 1)], pmf, use_line_collection=True)
d2l.plt.xlabel('x')
d2l.plt.ylabel('p.m.f.')
d2l.plt.show()
```

Now, let's plot the cumulative distribution function (A.10).

```
x = torch.arange(-1, 11, 0.01)
cmf = torch.cumsum(pmf, dim=0)

def F(x):
    return 0 if x < 0 else 1 if x > n else cmf[int(x)]

d2l.plot(x, torch.tensor([F(y) for y in x.tolist()]), 'x', 'c.d.f.')
```

If $X \sim \textrm{Binomial}(n, p)$, then:

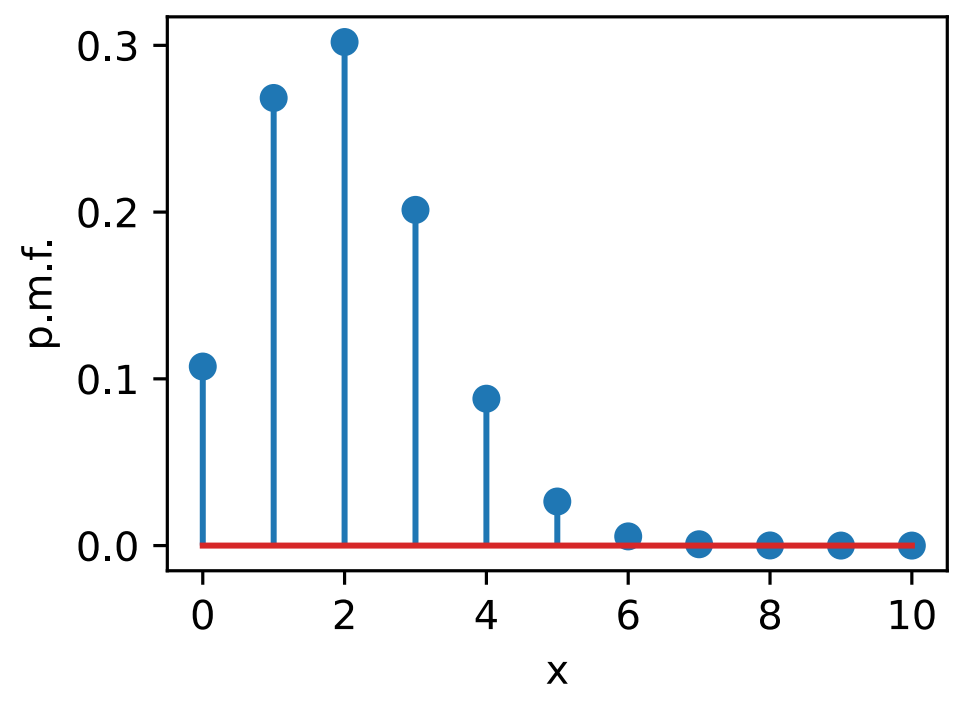

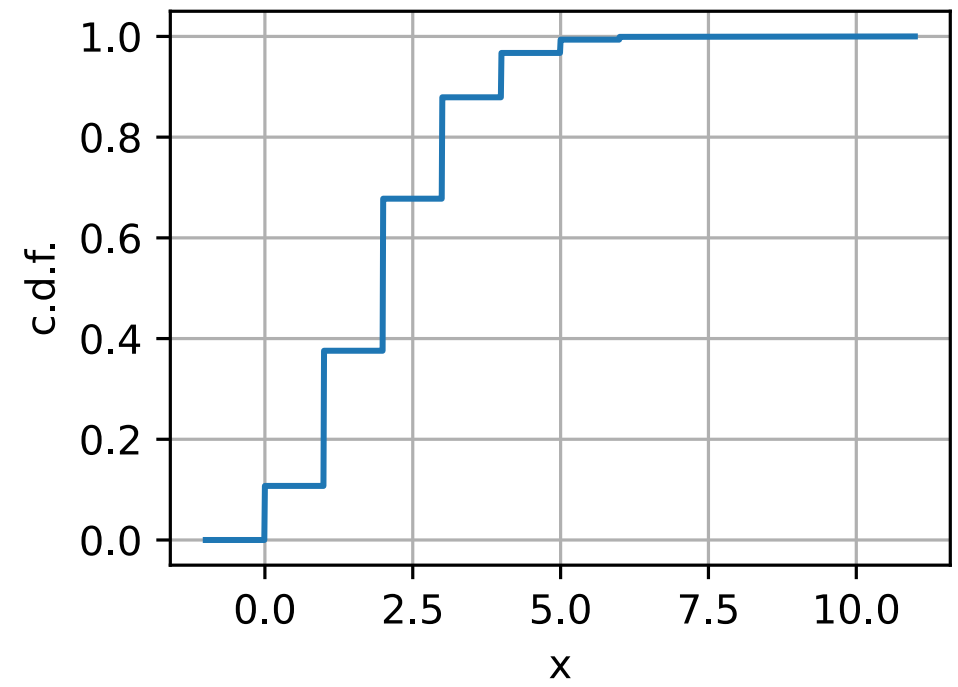

- $\mu_X = np$,
- $\sigma_X^2 = np(1-p)$.

This follows from the linearity of expected value over the sum of $n$ Bernoulli random variables, and the fact that the variance of the sum of independent random variables is the sum of the variances. This can be sampled as follows.

```
m = torch.distributions.binomial.Binomial(n, p)
m.sample(sample_shape=(10, 10))
```

```
tensor([[3., 3., 3., 3., 0., 2., 1., 3., 2., 1.],
        [4., 0., 0., 4., 2., 1., 0., 3., 0., 1.],
        [3., 2., 3., 0., 1., 2., 3., 1., 1., 2.],
        [4., 1., 1., 4., 3., 2., 1., 2., 1., 3.],
        [2., 2., 1., 1., 3., 3., 2., 1., 2., 2.],
        [3., 1., 4., 5., 2., 1., 2., 2., 0., 0.],
        [3., 2., 0., 2., 0., 1., 2., 5., 2., 1.],
        [2., 2., 5., 0., 1., 2., 1., 4., 3., 4.],
        [0., 3., 2., 0., 1., 0., 2., 2., 1., 3.],
        [2., 3., 1., 1., 1., 2., 2., 2., 2., 2.]])
```

## A.8.5 Poisson

---

Let's now perform a thought experiment. We are standing at a bus stop and we want to know how many buses will arrive in the next minute. Let's start by considering $X^{(1)} \sim \text{Bernoulli}(p)$ which is simply the probability that a bus arrives in the one minute window. For bus stops far from an urban center, this might be a pretty good approximation. We may never see more than one bus in a minute.

However, if we are in a busy area, it is possible or even likely that two buses will arrive. We can model this by splitting our random variable into two parts for the first 30 seconds, or the second 30 seconds. In this case we can write

$$X^{(2)} \sim X^{(2)}_1 + X^{(2)}_2, \tag{A.11}$$

where $X^{(2)}$ is the total sum, and $X^{(2)}_i \sim \text{Bernoulli}(p/2)$. The total distribution is then $X^{(2)} \sim \text{Binomial}(2, p/2)$.

Why stop here? Let's continue to split that minute into $n$ parts. By the same reasoning as above, we see that

$$X^{(n)} \sim \text{Binomial}(n, p/n). \tag{A.12}$$

Consider these random variables. By the previous section, we know that (A.12) has mean $\mu_{X^{(n)}} = n(p/n) = p$, and variance $\sigma^2_{X^{(n)}} = n(p/n)(1-(p/n)) = p(1-p/n)$. If we take $n \rightarrow \infty$, we can see that these numbers stabilize to $\mu_{X^{(\infty)}} = p$, and variance $\sigma^2_{X^{(\infty)}} = p$. This indicates that there *could be* some random variable we can define in this infinite subdivision limit.

This should not come as too much of a surprise, since in the real world we can just count the number of bus arrivals, however it is nice to see that our mathematical model is well defined. This discussion can be made formal as the *law of rare events*.

Following through this reasoning carefully, we can arrive at the following model. We will say that $X \sim \text{Poisson}(\lambda)$ if it is a random variable which takes the values $\{0, 1, 2, \ldots\}$ with probability

$$p_k = \frac{\lambda^k e^{-\lambda}}{k!}. \tag{A.13}$$

The value $\lambda > 0$ is known as the *rate* (or the *shape* parameter), and denotes the average number of arrivals we expect in one unit of time.

We may sum this probability mass function to get the cumulative distribution function.

$$F(x) = \begin{cases} 0 & x < 0, \\ e^{-\lambda}\sum_{m=0}^{k} \frac{\lambda^m}{m!} & k \leq x < k+1 \text{ with } 0 \leq k. \end{cases} \tag{A.14}$$

Let's first plot the probability mass function (A.13).

```
lam = 5.0

xs = [i for i in range(20)]
```

```
pmf = torch.tensor([torch.exp(torch.tensor(-lam)) * lam**k
                    / factorial(k) for k in xs])

d2l.plt.stem(xs, pmf, use_line_collection=True)
d2l.plt.xlabel('x')
d2l.plt.ylabel('p.m.f.')
d2l.plt.show()
```

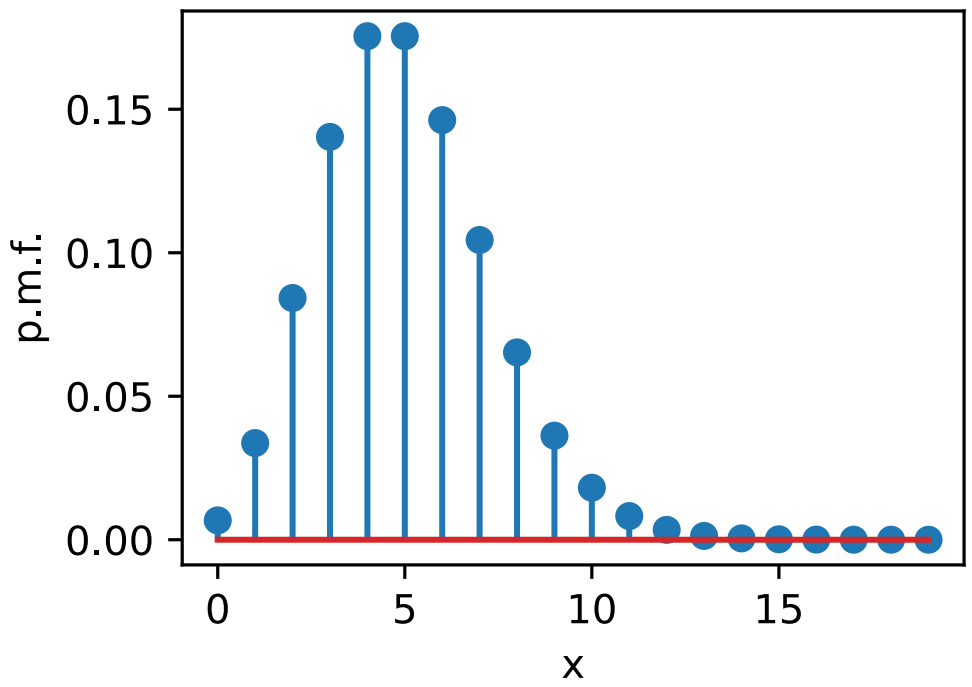

Now, let's plot the cumulative distribution function (A.14).

```
x = torch.arange(-1, 21, 0.01)
cmf = torch.cumsum(pmf, dim=0)
def F(x):
    return 0 if x < 0 else 1 if x > n else cmf[int(x)]

d2l.plot(x, torch.tensor([F(y) for y in x.tolist()]), 'x', 'c.d.f.')
```

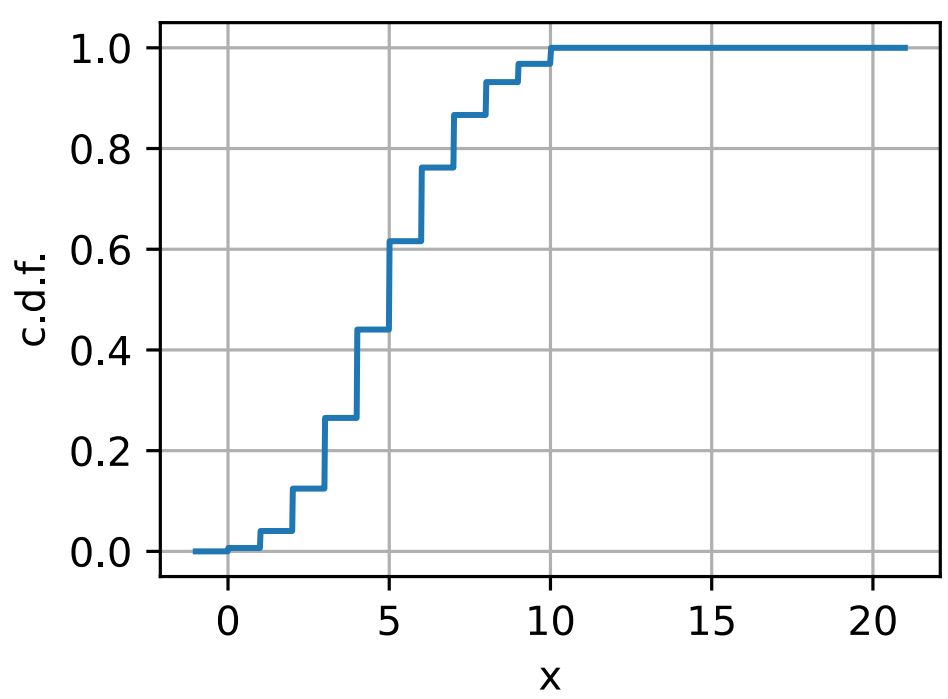

As we saw above, the means and variances are particularly concise. If $X \sim \text{Poisson}(\lambda)$, then:

- $\mu_X = \lambda$,
- $\sigma_X^2 = \lambda$.

This can be sampled as follows.

```
m = torch.distributions.poisson.Poisson(lam)
m.sample((10, 10))
```

```
tensor([[ 7.,  2.,  1.,  6.,  2., 10., 10.,  6.,  2.,  4.],
        [ 5.,  2.,  6.,  4.,  2.,  4.,  5.,  3.,  2.,  9.],
        [ 3.,  6.,  6.,  5.,  3.,  9., 12.,  7.,  4.,  6.],
        [ 8., 10.,  9.,  4.,  4.,  7.,  5.,  4.,  3.,  8.],
        [ 2.,  2., 10.,  6.,  7.,  5.,  9.,  1.,  7.,  7.],
        [ 6.,  4.,  3.,  8.,  4.,  2.,  3.,  5.,  3.,  5.],
        [ 0.,  3.,  3.,  4.,  7.,  6.,  2.,  3.,  4.,  4.],
        [ 8.,  5.,  4.,  2.,  5.,  5.,  9.,  7.,  8.,  4.],
        [ 5.,  4.,  7.,  1.,  2., 11.,  2.,  9.,  3.,  5.],
        [ 2.,  9.,  7.,  8.,  8.,  6.,  1.,  2.,  6.,  7.]])
```

### A.8.6 Gaussian

Now Let's try a different, but related experiment. Let's say we again are performing $n$ independent Bernoulli$(p)$ measurements $X_i$. The distribution of the sum of these is $X^{(n)} \sim$ Binomial$(n, p)$. Rather than taking a limit as $n$ increases and $p$ decreases, Let's fix $p$, and then send $n \to \infty$. In this case $\mu_{X^{(n)}} = np \to \infty$ and $\sigma^2_{X^{(n)}} = np(1-p) \to \infty$, so there is no reason to think this limit should be well defined.

However, not all hope is lost! Let's just make the mean and variance be well behaved by defining

$$Y^{(n)} = \frac{X^{(n)} - \mu_{X^{(n)}}}{\sigma_{X^{(n)}}}. \tag{A.15}$$

This can be seen to have mean zero and variance one, and so it is plausible to believe that it will converge to some limiting distribution. If we plot what these distributions look like, we will become even more convinced that it will work.

```
p = 0.2
ns = [1, 10, 100, 1000]
d2l.plt.figure(figsize=(10, 3))
for i in range(4):
    n = ns[i]
    pmf = torch.tensor([p**i * (1-p)**(n-i) * binom(n, i)
                        for i in range(n + 1)])
    d2l.plt.subplot(1, 4, i + 1)
    d2l.plt.stem([(i - n*p)/torch.sqrt(torch.tensor(n*p*(1 - p)))
                  for i in range(n + 1)], pmf,
                 use_line_collection=True)
    d2l.plt.xlim([-4, 4])
    d2l.plt.xlabel('x')
    d2l.plt.ylabel('p.m.f.')
    d2l.plt.title("n = {}".format(n))
d2l.plt.show()
```

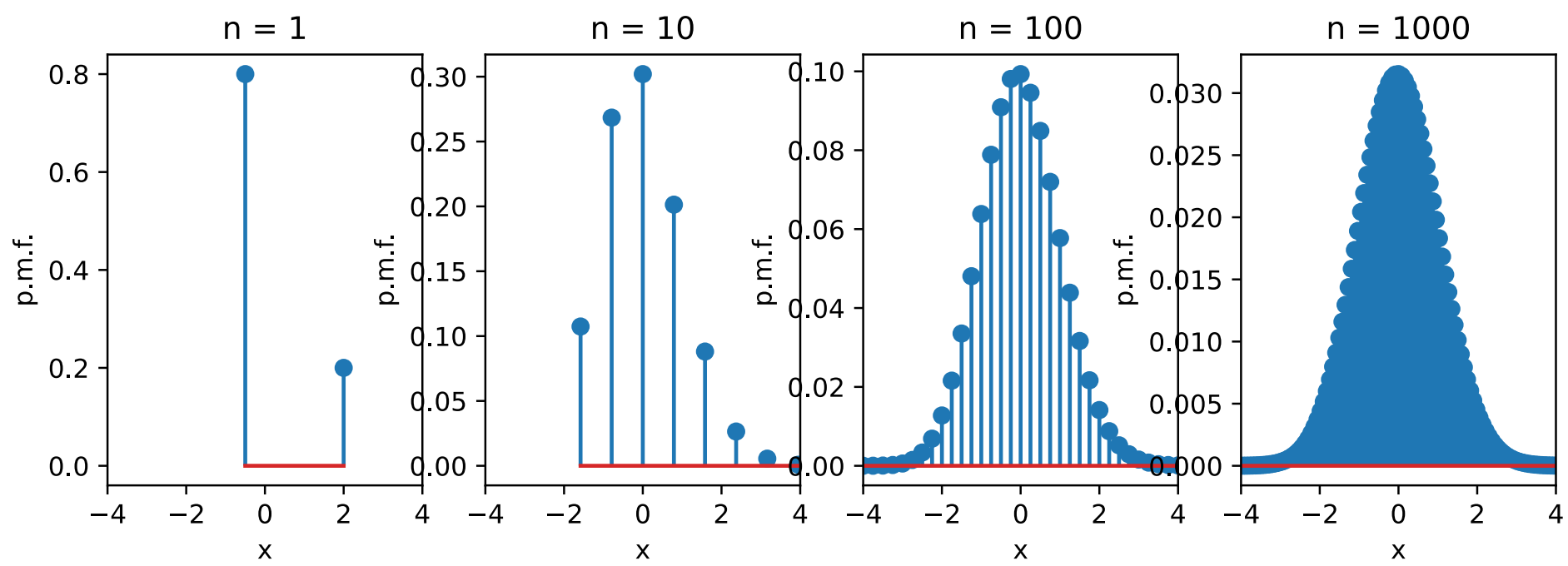

One thing to note: compared to the Poisson case, we are now dividing by the standard deviation which means that we are squeezing the possible outcomes into smaller and smaller areas. This is an indication that our limit will no longer be discrete, but rather continuous.

A derivation of what occurs is beyond the scope of this document, but the *central limit theorem* states that as $n \to \infty$, this will yield the Gaussian Distribution (or sometimes normal distribution). More explicitly, for any $a, b$:

$$\lim_{n\to\infty} P(Y^{(n)} \in [a, b]) = P(\mathcal{N}(0, 1) \in [a, b]), \tag{A.16}$$

where we say a random variable is normally distributed with given mean $\mu$ and variance $\sigma^2$, written $X \sim \mathcal{N}(\mu, \sigma^2)$ if $X$ has density

$$p_X(x) = \frac{1}{\sqrt{2\pi\sigma^2}} e^{-\frac{(x-\mu)^2}{2\sigma^2}}. \tag{A.17}$$

Let's first plot the probability density function (A.17).

```
mu, sigma = 0, 1

x = torch.arange(-3, 3, 0.01)
p = 1 / torch.sqrt(2 * torch.pi * sigma**2) * torch.exp(
    -(x - mu)**2 / (2 * sigma**2))

d2l.plot(x, p, 'x', 'p.d.f.')
```

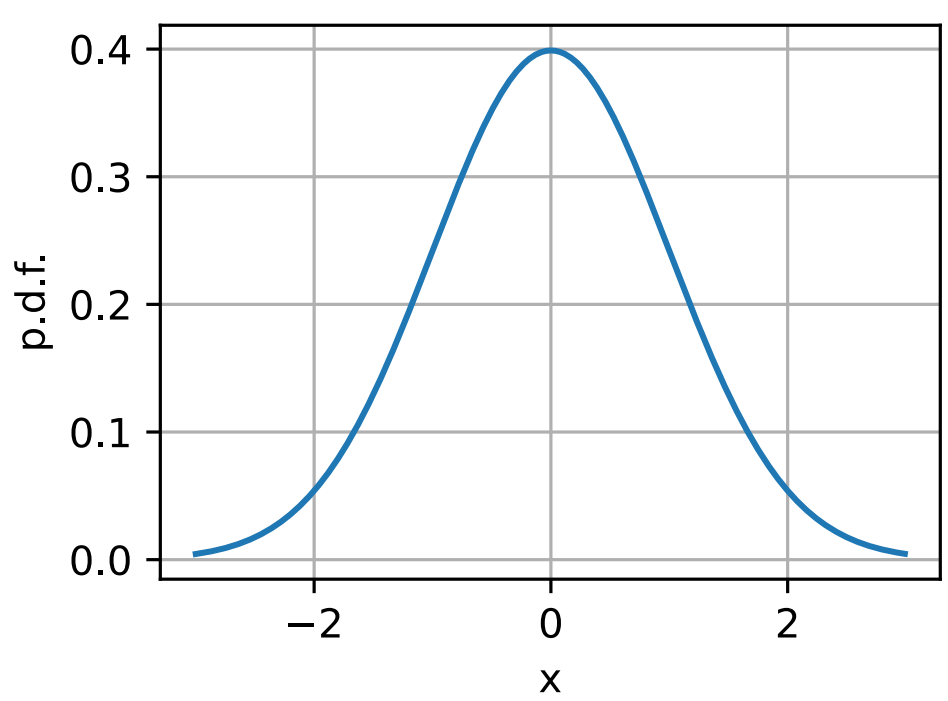

Now, let's plot the cumulative distribution function. It is beyond the scope of this appendix, but the Gaussian c.d.f. does not have a closed-form formula in terms of more elementary functions. We will use `erf` which provides a way to compute this integral numerically.

```
def phi(x):
    return (1.0 + erf((x - mu) / (sigma * torch.sqrt(torch.tensor(2.))))) / 2.0

d2l.plot(x, torch.tensor([phi(y) for y in x.tolist()]), 'x', 'c.d.f.')
```

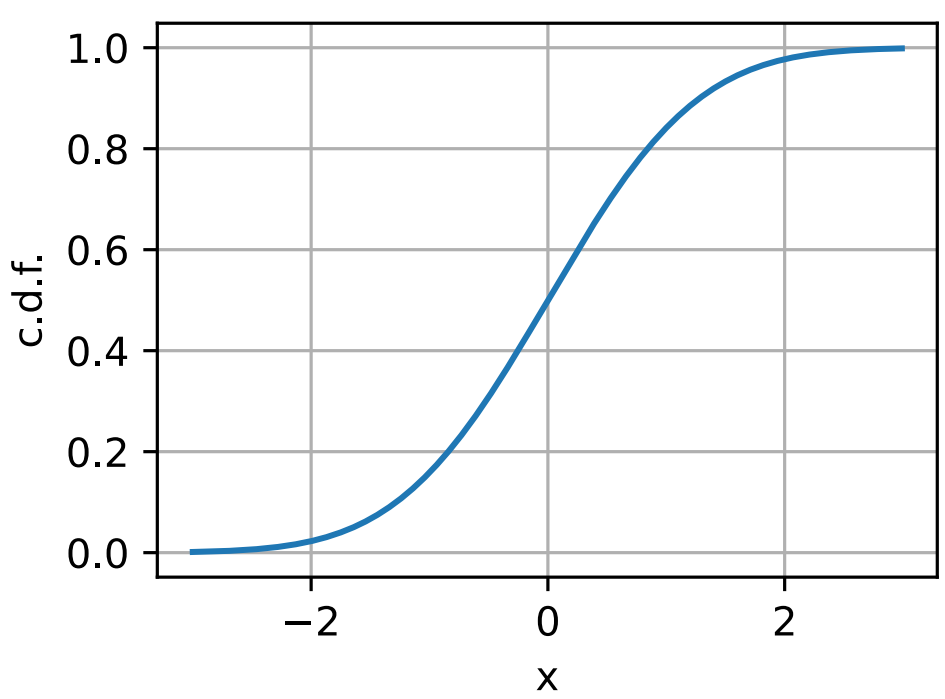

Keen-eyed readers will recognize some of these terms. Indeed, we encountered this integral in Section A.5. Indeed we need exactly that computation to see that this $p_X(x)$ has total area one and is thus a valid density.

Our choice of working with coin flips made computations shorter, but nothing about that choice was fundamental. Indeed, if we take any collection of independent identically distributed random variables $X_i$, and form

$$X^{(N)} = \sum_{i=1}^{N} X_i. \tag{A.18}$$

Then

$$\frac{X^{(N)} - \mu_{X^{(N)}}}{\sigma_{X^{(N)}}} \tag{A.19}$$

will be approximately Gaussian. There are additional requirements needed to make it work, most commonly $E[X^4] < \infty$, but the philosophy is clear.

The central limit theorem is the reason why the Gaussian is fundamental to probability, statistics, and machine learning. Whenever we can say that something we measured is a sum of many small independent contributions, we can assume that the thing being measured will be close to Gaussian.

There are many more fascinating properties of Gaussians, and we would like to discuss one more here. The Gaussian is what is known as a *maximum entropy distribution*. We will get into entropy more deeply in Section A.11, however all we need to know at this point is that it is a measure of randomness. In a rigorous mathematical sense, we can think of the Gaussian as the *most* random choice of random variable with fixed mean and variance. Thus, if we know that our random variable has some mean and variance, the Gaussian is in a sense the most conservative choice of distribution we can make.

To close the section, let's recall that if $X \sim \mathcal{N}(\mu, \sigma^2)$, then:

- $\mu_X = \mu$,
- $\sigma_X^2 = \sigma^2$.

We can sample from the Gaussian (or standard normal) distribution as shown below.

```
torch.normal(mu, sigma, size=(10, 10))
```

```
tensor([[ 0.1100,  1.2204, -0.0183, -0.0438,  0.1343, -1.7406, -1.0956, -1.
↪0103,
         -0.8703,  0.9853],
        [-1.7443, -0.8289,  1.8382, -1.2046,  0.4454, -0.1467, -1.2068, -1.
↪1904,
          0.7402, -0.9666],
        [ 0.4668, -0.5373, -0.6894, -0.5451,  0.5279,  1.7936, -0.3843,  1.
↪6206,
          0.3415, -0.6788],
        [-0.4203,  1.5654,  0.5024,  0.5272,  2.4095, -1.4125,  0.0776, -0.
↪7675,
         -2.1896, -0.4801],
        [-0.4477,  1.2597, -1.7383,  0.2176,  2.0329, -0.0337, -0.3814,  1.
↪8359,
          0.3762, -1.0739],
        [-0.5160,  2.0706, -0.0820,  0.5688, -0.3288,  1.8007,  0.4799, -0.
↪6264,
          0.3517, -0.4274],
        [-1.1953,  1.4043, -0.8325, -0.2550,  0.2850,  0.0788, -1.6121, -1.
↪0295,
         -0.5723, -0.5142],
        [-0.0106, -0.6287,  0.6475,  1.5597,  0.3415,  0.8483,  1.5621,  0.
↪0976,
          1.8702, -1.8762],
        [-1.0095, -0.2200,  0.1529, -0.3789,  0.1588,  0.6434,  0.5891,  2.
↪1004,
         -1.4606, -0.9920],
        [-0.2421,  0.4331,  0.5397, -1.2819,  0.7858, -0.5514,  2.0629,  0.
↪3418,
         -2.1320, -0.5923]])
```

## A.8.7 Exponential Family

One shared property for all the distributions listed above is that they all belong to which is known as the *exponential family*. The exponential family is a set of distributions whose density can be expressed in the following form:

$$p(\mathbf{x} \mid \boldsymbol{\eta}) = h(\mathbf{x}) \cdot \exp\left(\boldsymbol{\eta}^\top \cdot T(\mathbf{x}) - A(\boldsymbol{\eta})\right) \tag{A.20}$$

As this definition can be a little subtle, let's examine it closely.

First, $h(\mathbf{x})$ is known as the *underlying measure* or the *base measure*. This can be viewed as an original choice of measure we are modifying with our exponential weight.

Second, we have the vector $\boldsymbol{\eta} = (\eta_1, \eta_2, ..., \eta_l) \in \mathbb{R}^l$ called the *natural parameters* or *canonical parameters*. These define how the base measure will be modified. The natural parameters enter into the new measure by taking the dot product of these parameters against some function $T(\cdot)$ of $\mathbf{x} = (x_1, x_2, ..., x_n) \in \mathbb{R}^n$ and exponentiated. The vector $T(\mathbf{x}) = (T_1(\mathbf{x}), T_2(\mathbf{x}), ..., T_l(\mathbf{x}))$ is called the *sufficient statistics* for $\boldsymbol{\eta}$. This name is used since the information represented by $T(\mathbf{x})$ is sufficient to calculate the probability density and no other information from the sample $\mathbf{x}$'s are required.

Third, we have $A(\boldsymbol{\eta})$, which is referred to as the *cumulant function*, which ensures that the above distribution (A.20) integrates to one, i.e.,

$$A(\boldsymbol{\eta}) = \log\left[\int h(\mathbf{x}) \cdot \exp\left(\boldsymbol{\eta}^\top \cdot T(\mathbf{x})\right) d\mathbf{x}\right]. \tag{A.21}$$

To be concrete, let's consider the Gaussian. Assuming that $\mathbf{x}$ is an univariate variable, we saw that it had a density of

$$\begin{aligned} p(x \mid \mu, \sigma) &= \frac{1}{\sqrt{2\pi\sigma^2}} \cdot \exp\left\{\frac{-(x-\mu)^2}{2\sigma^2}\right\} \\ &= \frac{1}{\sqrt{2\pi}} \cdot \exp\left\{\frac{\mu}{\sigma^2}x - \frac{1}{2\sigma^2}x^2 - \left(\frac{1}{2\sigma^2}\mu^2 + \log(\sigma)\right)\right\}. \end{aligned} \tag{A.22}$$

This matches the definition of the exponential family with:

- *underlying measure*: $h(x) = \frac{1}{\sqrt{2\pi}}$,
- *natural parameters*: $\boldsymbol{\eta} = \begin{bmatrix}\eta_1 \\ \eta_2\end{bmatrix} = \begin{bmatrix}\frac{\mu}{\sigma^2} \\ \frac{1}{2\sigma^2}\end{bmatrix}$,
- *sufficient statistics*: $T(x) = \begin{bmatrix} x \\ -x^2\end{bmatrix}$, and
- *cumulant function*: $A(\boldsymbol{\eta}) = \frac{1}{2\sigma^2}\mu^2 + \log(\sigma) = \frac{\eta_1^2}{4\eta_2} - \frac{1}{2}\log(2\eta_2)$.

It is worth noting that the exact choice of each of above terms is somewhat arbitrary. Indeed, the important feature is that the distribution can be expressed in this form, not the exact form itself.

As we allude to in Section 4.1.2, a widely used technique is to assume that the final output $\mathbf{y}$ follows an exponential family distribution. The exponential family is a common and powerful family of distributions encountered frequently in machine learning.

## A.8.8 Summary

- Bernoulli random variables can be used to model events with a yes/no outcome.

- Discrete uniform distributions model selects from a finite set of possibilities.
- Continuous uniform distributions select from an interval.
- Binomial distributions model a series of Bernoulli random variables, and count the number of successes.
- Poisson random variables model the arrival of rare events.
- Gaussian random variables model the result of adding a large number of independent random variables together.
- All the above distributions belong to exponential family.

### A.8.9 Exercises

1. What is the standard deviation of a random variable that is the difference $X - Y$ of two independent binomial random variables $X, Y \sim \mathrm{Binomial}(16, 1/2)$.
2. If we take a Poisson random variable $X \sim \mathrm{Poisson}(\lambda)$ and consider $(X - \lambda)/\sqrt{\lambda}$ as $\lambda \to \infty$, we can show that this becomes approximately Gaussian. Why does this make sense?
3. What is the probability mass function for a sum of two discrete uniform random variables on $n$ elements?

Discussions[285].

## A.9 Naive Bayes

Throughout the previous sections, we learned about the theory of probability and random variables. To put this theory to work, let's introduce the *naive Bayes* classifier. This uses nothing but probabilistic fundamentals to allow us to perform classification of digits.

Learning is all about making assumptions. If we want to classify a new data example that we have never seen before we have to make some assumptions about which data examples are similar to each other. The naive Bayes classifier, a popular and remarkably clear algorithm, assumes all features are independent from each other to simplify the computation. In this section, we will apply this model to recognize characters in images.

```
%matplotlib inline
import math
import torch
import torchvision
from d2l import torch as d2l

d2l.use_svg_display()
```

### A.9.1 Optical Character Recognition

MNIST (LeCun *et al.*, 1998) is one of widely used datasets. It contains 60,000 images for training and 10,000 images for validation. Each image contains a handwritten digit from 0 to 9. The task is classifying each image into the corresponding digit.

Gluon provides a `MNIST` class in the `data.vision` module to automatically retrieve the dataset from the Internet. Subsequently, Gluon will use the already-downloaded local copy. We specify whether we are requesting the training set or the test set by setting the value of the parameter `train` to `True` or `False`, respectively. Each image is a grayscale image with both width and height of 28 with shape (28,28,1). We use a customized transformation to remove the last channel dimension. In addition, the dataset represents each pixel by an unsigned 8-bit integer. We quantize them into binary features to simplify the problem.

```
data_transform = torchvision.transforms.Compose([
    torchvision.transforms.ToTensor(),
    lambda x: torch.floor(x * 255 / 128).squeeze(dim=0)
])

mnist_train = torchvision.datasets.MNIST(
    root='./temp', train=True, transform=data_transform, download=True)
mnist_test = torchvision.datasets.MNIST(
    root='./temp', train=False, transform=data_transform, download=True)
```

We can access a particular example, which contains the image and the corresponding label.

```
image, label = mnist_train[2]
image.shape, label
```

```
(torch.Size([28, 28]), 4)
```

Our example, stored here in the variable `image`, corresponds to an image with a height and width of 28 pixels.

```
image.shape, image.dtype
```

```
(torch.Size([28, 28]), torch.float32)
```

Our code stores the label of each image as a scalar. Its type is a 32-bit integer.

```
label, type(label)
```

```
(4, int)
```

We can also access multiple examples at the same time.

```
images = torch.stack([mnist_train[i][0] for i in range(10, 38)], dim=0)
labels = torch.tensor([mnist_train[i][1] for i in range(10, 38)])
images.shape, labels.shape
```

```
(torch.Size([28, 28, 28]), torch.Size([28]))
```

Let's visualize these examples.

```
d2l.show_images(images, 2, 9);
```

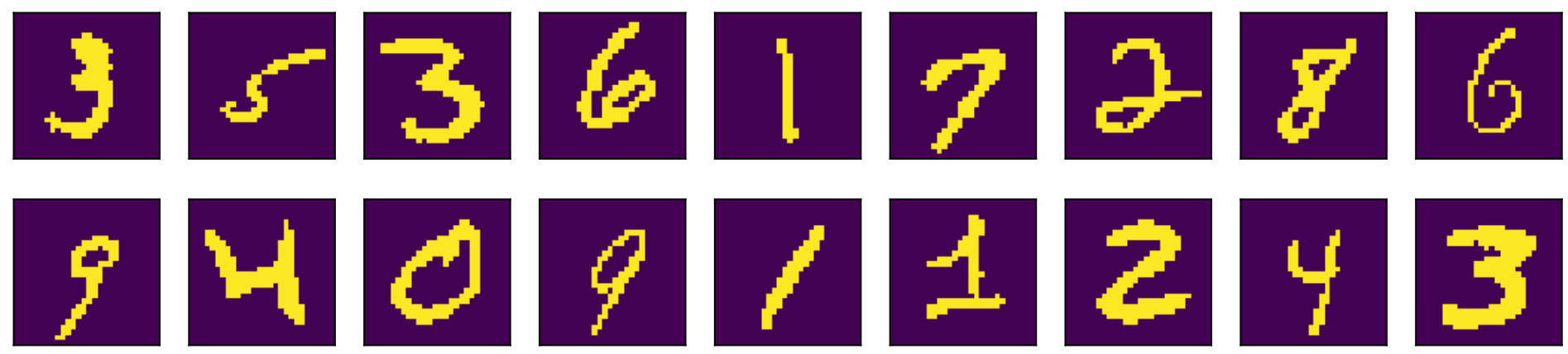

## A.9.2 The Probabilistic Model for Classification

In a classification task, we map an example into a category. Here an example is a grayscale $28\times28$ image, and a category is a digit. (Refer to Section 4.1 for a more detailed explanation.) One natural way to express the classification task is via the probabilistic question: what is the most likely label given the features (i.e., image pixels)? Denote by $\mathbf{x} \in \mathbb{R}^d$ the features of the example and $y \in \mathbb{R}$ the label. Here features are image pixels, where we can reshape a 2-dimensional image to a vector so that $d = 28^2 = 784$, and labels are digits. The probability of the label given the features is $p(y \mid \mathbf{x})$. If we are able to compute these probabilities, which are $p(y \mid \mathbf{x})$ for $y = 0, \ldots, 9$ in our example, then the classifier will output the prediction $\hat{y}$ given by the expression:

$$\hat{y} = \mathrm{argmax} \> p(y \mid \mathbf{x}). \tag{A.1}$$

Unfortunately, this requires that we estimate $p(y \mid \mathbf{x})$ for every value of $\mathbf{x} = x_1, ..., x_d$. Imagine that each feature could take one of 2 values. For example, the feature $x_1 = 1$ might signify that the word apple appears in a given document and $x_1 = 0$ would signify that it does not. If we had 30 such binary features, that would mean that we need to be prepared to classify any of $2^{30}$ (over 1 billion!) possible values of the input vector $\mathbf{x}$.

Moreover, where is the learning? If we need to see every single possible example in order to predict the corresponding label then we are not really learning a pattern but just memorizing the dataset.

## A.9.3 The Naive Bayes Classifier

Fortunately, by making some assumptions about conditional independence, we can introduce some inductive bias and build a model capable of generalizing from a comparatively modest selection of training examples. To begin, let's use Bayes theorem, to express the classifier as

$$\hat{y} = \mathrm{argmax}_y \, p(y \mid \mathbf{x}) = \mathrm{argmax}_y \, \frac{p(\mathbf{x} \mid y) p(y)}{p(\mathbf{x})}. \tag{A.2}$$

Note that the denominator is the normalizing term $p(\mathbf{x})$ which does not depend on the value of the label $y$. As a result, we only need to worry about comparing the numerator across different values of $y$. Even if calculating the denominator turned out to be intractable, we could get away with ignoring it, so long as we could evaluate the numerator. Fortunately, even if we wanted to recover the normalizing constant, we could. We can always recover the normalization term since $\sum_y p(y \mid \mathbf{x}) = 1$.

Now, let's focus on $p(\mathbf{x} \mid y)$. Using the chain rule of probability, we can express the term $p(\mathbf{x} \mid y)$ as

$$p(x_1 \mid y) \cdot p(x_2 \mid x_1, y) \cdot \ldots \cdot p(x_d \mid x_1, \ldots, x_{d-1}, y). \tag{A.3}$$

By itself, this expression does not get us any further. We still must estimate roughly $2^d$ parameters. However, if we assume that *the features are conditionally independent of each other, given the label*, then suddenly we are in much better shape, as this term simplifies to $\prod_i p(x_i \mid y)$, giving us the predictor

$$\hat{y} = \mathrm{argmax}_y \prod_{i=1}^{d} p(x_i \mid y) p(y). \tag{A.4}$$

If we can estimate $p(x_i = 1 \mid y)$ for every $i$ and $y$, and save its value in $P_{xy}[i, y]$, here $P_{xy}$ is a $d \times n$ matrix with $n$ being the number of classes and $y \in \{1, \ldots, n\}$, then we can also use this to estimate $p(x_i = 0 \mid y)$, i.e.,

$$p(x_i = t_i \mid y) = \begin{cases} P_{xy}[i, y] & \text{for } t_i = 1; \\ 1 - P_{xy}[i, y] & \text{for } t_i = 0. \end{cases} \tag{A.5}$$

In addition, we estimate $p(y)$ for every $y$ and save it in $P_y[y]$, with $P_y$ a $n$-length vector. Then, for any new example $\mathbf{t} = (t_1, t_2, \ldots, t_d)$, we could compute

$$\begin{aligned} \hat{y} &= \mathrm{argmax}_y \, p(y) \prod_{i=1}^{d} p(x_t = t_i \mid y) \\ &= \mathrm{argmax}_y \, P_y[y] \prod_{i=1}^{d} P_{xy}[i, y]^{t_i} \left(1 - P_{xy}[i, y]\right)^{1-t_i} \end{aligned} \tag{A.6}$$

for any $y$. So our assumption of conditional independence has taken the complexity of our model from an exponential dependence on the number of features $O(2^d n)$ to a linear dependence, which is $O(dn)$.

### A.9.4 Training

The problem now is that we do not know $P_{xy}$ and $P_y$. So we need to estimate their values given some training data first. This is *training* the model. Estimating $P_y$ is not too hard. Since we are only dealing with 10 classes, we may count the number of occurrences $n_y$ for each of the digits and divide it by the total amount of data $n$. For instance, if digit 8 occurs $n_8 = 5,800$ times and we have a total of $n = 60,000$ images, the probability estimate is $p(y = 8) = 0.0967$.

```
X = torch.stack([mnist_train[i][0] for i in range(len(mnist_train))], dim=0)
Y = torch.tensor([mnist_train[i][1] for i in range(len(mnist_train))])

n_y = torch.zeros(10)
for y in range(10):
    n_y[y] = (Y == y).sum()
P_y = n_y / n_y.sum()
P_y
```

```
tensor([0.0987, 0.1124, 0.0993, 0.1022, 0.0974, 0.0904, 0.0986, 0.1044, 0.0975,
        0.0992])
```

Now on to slightly more difficult things $P_{xy}$. Since we picked black and white images, $p(x_i \mid y)$ denotes the probability that pixel $i$ is switched on for class $y$. Just like before we can go and count the number of times $n_{iy}$ such that an event occurs and divide it by the total number of occurrences of $y$, i.e., $n_y$. But there is something slightly troubling: certain pixels may never be black (e.g., for well cropped images the corner pixels might always be white). A convenient way for statisticians to deal with this problem is to add pseudo counts to all occurrences. Hence, rather than $n_{iy}$ we use $n_{iy} + 1$ and instead of $n_y$ we use $n_y + 2$ (since there are two possible values pixel $i$ can take - it can either be black or white). This is also called *Laplace Smoothing*. It may seem ad-hoc, however it can be motivated from a Bayesian point-of-view by a Beta-binomial model.

```
n_x = torch.zeros((10, 28, 28))
for y in range(10):
    n_x[y] = torch.tensor(X.numpy()[Y.numpy() == y].sum(axis=0))
P_xy = (n_x + 1) / (n_y + 2).reshape(10, 1, 1)

d2l.show_images(P_xy, 2, 5);
```

By visualizing these $10 \times 28 \times 28$ probabilities (for each pixel for each class) we could get some mean looking digits.

Now we can use (A.6) to predict a new image. Given $\mathbf{x}$, the following functions computes $p(\mathbf{x} \mid y)p(y)$ for every $y$.

```
def bayes_pred(x):
    x = x.unsqueeze(0)  # (28, 28) -> (1, 28, 28)
```

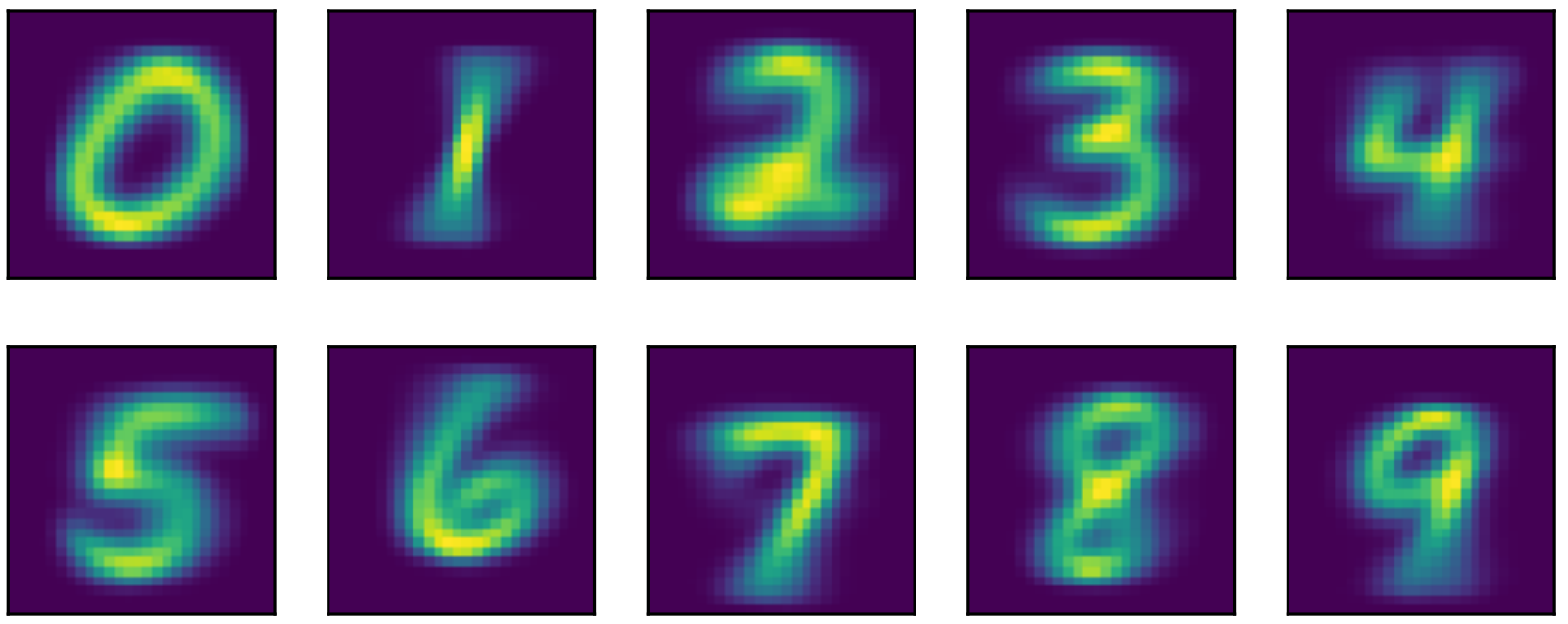

```
    p_xy = P_xy * x + (1 - P_xy)*(1 - x)
    p_xy = p_xy.reshape(10, -1).prod(dim=1)  # p(x|y)
    return p_xy * P_y

image, label = mnist_test[0]
bayes_pred(image)
```

```
tensor([0., 0., 0., 0., 0., 0., 0., 0., 0., 0.])
```

This went horribly wrong! To find out why, let's look at the per pixel probabilities. They are typically numbers between $0.001$ and $1$. We are multiplying $784$ of them. At this point it is worth mentioning that we are calculating these numbers on a computer, hence with a fixed range for the exponent. What happens is that we experience *numerical underflow*, i.e., multiplying all the small numbers leads to something even smaller until it is rounded down to zero. We discussed this as a theoretical issue in Section A.7, but we see the phenomena clearly here in practice.

As discussed in that section, we fix this by use the fact that $\log ab = \log a + \log b$, i.e., we switch to summing logarithms. Even if both $a$ and $b$ are small numbers, the logarithm values should be in a proper range.

```
a = 0.1
print('underflow:', a**784)
print('logarithm is normal:', 784*math.log(a))
```

```
underflow: 0.0
logarithm is normal: -1805.2267129073316
```

Since the logarithm is an increasing function, we can rewrite (A.6) as

$$\hat{y} = \operatorname{argmax}_y \ \log P_y[y] + \sum_{i=1}^{d} \Big[ t_i \log P_{xy}[x_i, y] + (1 - t_i) \log(1 - P_{xy}[x_i, y]) \Big]. \tag{A.7}$$

We can implement the following stable version:

```
log_P_xy = torch.log(P_xy)
log_P_xy_neg = torch.log(1 - P_xy)
log_P_y = torch.log(P_y)

def bayes_pred_stable(x):
    x = x.unsqueeze(0)  # (28, 28) -> (1, 28, 28)
    p_xy = log_P_xy * x + log_P_xy_neg * (1 - x)
    p_xy = p_xy.reshape(10, -1).sum(axis=1)  # p(x|y)
    return p_xy + log_P_y

py = bayes_pred_stable(image)
py
```

```
tensor([-268.9725, -301.7044, -245.1951, -218.8738, -193.4570, -206.0909,
        -292.5226, -114.6257, -220.3313, -163.1784])
```

We may now check if the prediction is correct.

```
py.argmax(dim=0) == label
```

```
tensor(True)
```

If we now predict a few validation examples, we can see the Bayes classifier works pretty well.

```
def predict(X):
    return [bayes_pred_stable(x).argmax(dim=0).type(torch.int32).item()
            for x in X]

X = torch.stack([mnist_test[i][0] for i in range(18)], dim=0)
y = torch.tensor([mnist_test[i][1] for i in range(18)])
preds = predict(X)
d2l.show_images(X, 2, 9, titles=[str(d) for d in preds]);
```

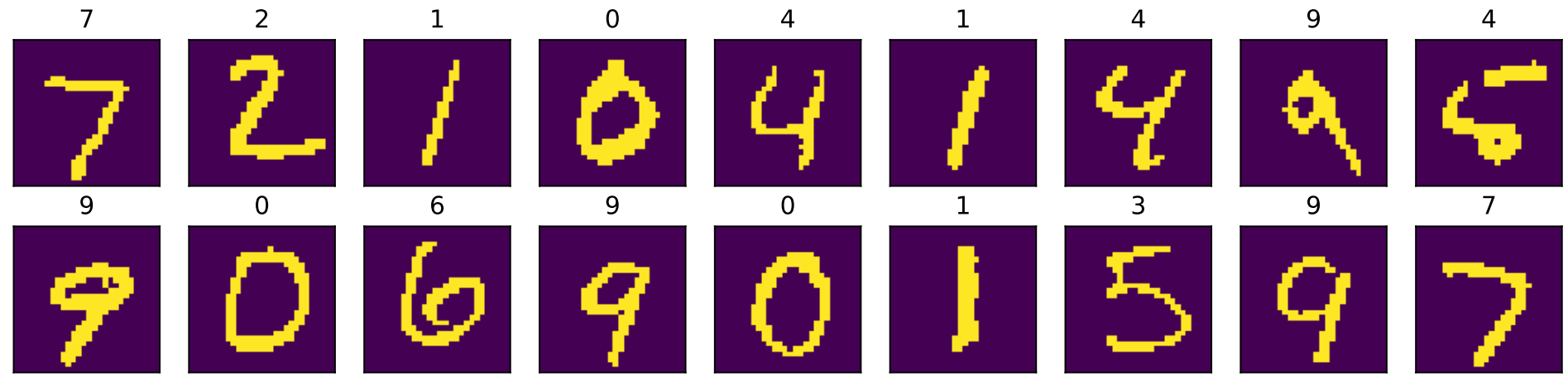

Finally, let's compute the overall accuracy of the classifier.

```
X = torch.stack([mnist_test[i][0] for i in range(len(mnist_test))], dim=0)
y = torch.tensor([mnist_test[i][1] for i in range(len(mnist_test))])
preds = torch.tensor(predict(X), dtype=torch.int32)
float((preds == y).sum()) / len(y)  # Validation accuracy
```

```
0.8427
```

Modern deep networks achieve error rates of less than $0.01$. The relatively poor performance is due to the incorrect statistical assumptions that we made in our model: we assumed that each and every pixel are *independently* generated, depending only on the label. This is clearly not how humans write digits, and this wrong assumption led to the downfall of our overly naive (Bayes) classifier.

### A.9.5 Summary

- Using Bayes' rule, a classifier can be made by assuming all observed features are independent.
- This classifier can be trained on a dataset by counting the number of occurrences of combinations of labels and pixel values.
- This classifier was the gold standard for decades for tasks such as spam detection.

### A.9.6 Exercises

1. Consider the dataset $[[0,0],[0,1],[1,0],[1,1]]$ with labels given by the XOR of the two elements $[0,1,1,0]$. What are the probabilities for a Naive Bayes classifier built on this dataset. Does it successfully classify our points? If not, what assumptions are violated?
2. Suppose that we did not use Laplace smoothing when estimating probabilities and a data example arrived at testing time which contained a value never observed in training. What would the model output?
3. The naive Bayes classifier is a specific example of a Bayesian network, where the dependence of random variables are encoded with a graph structure. While the full theory is beyond the scope of this section (see Koller and Friedman (2009) for full details), explain why allowing explicit dependence between the two input variables in the XOR model allows for the creation of a successful classifier.

Discussions[286].

## A.10 Statistics

Undoubtedly, to be a top deep learning practitioner, the ability to train the state-of-the-art and high accurate models is crucial. However, it is often unclear when improvements are significant, or only the result of random fluctuations in the training process. To be able to discuss uncertainty in estimated values, we must learn some statistics.

The earliest reference of *statistics* can be traced back to an Arab scholar Al-Kindi in the $9^{\text{th}}$-century, who gave a detailed description of how to use statistics and frequency analysis to decipher encrypted messages. After 800 years, the modern statistics arose from Germany in 1700s, when the researchers focused on the demographic and economic data collection and analysis. Today, statistics is the science subject that concerns the collection, processing, analysis, interpretation and visualization of data. What is more, the core theory of statistics has been widely used in the research within academia, industry, and government.

More specifically, statistics can be divided to *descriptive statistics* and *statistical inference*. The former focus on summarizing and illustrating the features of a collection of observed data, which is referred to as a *sample*. The sample is drawn from a *population*, denotes the total set of similar individuals, items, or events of our experiment interests. Contrary to descriptive statistics, *statistical inference* further deduces the characteristics of a population from the given *samples*, based on the assumptions that the sample distribution can replicate the population distribution at some degree.

You may wonder: "What is the essential difference between machine learning and statistics?" Fundamentally speaking, statistics focuses on the inference problem. This type of problems includes modeling the relationship between the variables, such as causal inference, and testing the statistically significance of model parameters, such as A/B testing. In contrast, machine learning emphasizes on making accurate predictions, without explicitly programming and understanding each parameter's functionality.

In this section, we will introduce three types of statistics inference methods: evaluating and comparing estimators, conducting hypothesis tests, and constructing confidence intervals. These methods can help us infer the characteristics of a given population, i.e., the true parameter $\theta$. For brevity, we assume that the true parameter $\theta$ of a given population is a scalar value. It is straightforward to extend to the case where $\theta$ is a vector or a tensor, thus we omit it in our discussion.

## A.10.1 Evaluating and Comparing Estimators

In statistics, an *estimator* is a function of given samples used to estimate the true parameter $\theta$. We will write $\hat{\theta}_n = \hat{f}(x_1, \ldots, x_n)$ for the estimate of $\theta$ after observing the samples $\{x_1, x_2, \ldots, x_n\}$.

We have seen simple examples of estimators before in section Section A.7. If you have a number of samples from a Bernoulli random variable, then the maximum likelihood estimate for the probability the random variable is one can be obtained by counting the number of ones observed and dividing by the total number of samples. Similarly, an exercise asked you to show that the maximum likelihood estimate of the mean of a Gaussian given a number of samples is given by the average value of all the samples. These estimators will almost never give the true value of the parameter, but ideally for a large number of samples the estimate will be close.

As an example, we show below the true density of a Gaussian random variable with mean zero and variance one, along with a collection samples from that Gaussian. We constructed the $y$ coordinate so every point is visible and the relationship to the original density is clearer.

```
import torch
from d2l import torch as d2l

torch.pi = torch.acos(torch.zeros(1)) * 2  #define pi in torch

# Sample datapoints and create y coordinate
epsilon = 0.1
torch.manual_seed(8675309)
xs = torch.randn(size=(300,))

ys = torch.tensor(
    [torch.sum(torch.exp(-(xs[:i] - xs[i])**2 / (2 * epsilon**2))\
               / torch.sqrt(2*torch.pi*epsilon**2)) / len(xs)\
     for i in range(len(xs))])

# Compute true density
xd = torch.arange(torch.min(xs), torch.max(xs), 0.01)
yd = torch.exp(-xd**2/2) / torch.sqrt(2 * torch.pi)

# Plot the results
d2l.plot(xd, yd, 'x', 'density')
d2l.plt.scatter(xs, ys)
d2l.plt.axvline(x=0)
d2l.plt.axvline(x=torch.mean(xs), linestyle='--', color='purple')
d2l.plt.title(f'sample mean: {float(torch.mean(xs).item()):.2f}')
d2l.plt.show()
```

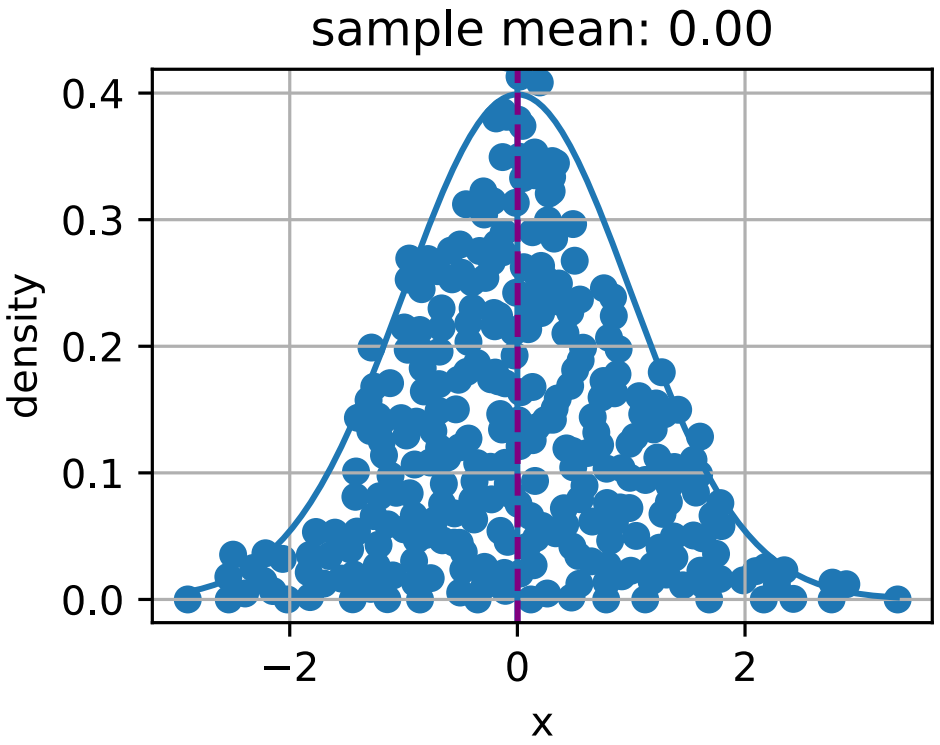

There can be many ways to compute an estimator of a parameter $\hat{\theta}_n$. In this section, we introduce three common methods to evaluate and compare estimators: the mean squared error, the standard deviation, and statistical bias.

### Mean Squared Error

Perhaps the simplest metric used to evaluate estimators is the *mean squared error (MSE)* (or $l_2$ loss) estimator which can be defined as

$$\text{MSE}(\hat{\theta}_n, \theta) = E[(\hat{\theta}_n - \theta)^2]. \tag{A.1}$$

This allows us to quantify the average squared deviation from the true value. MSE is always non-negative. If you have read Section 3.1, you will recognize it as the most commonly used regression loss function. As a measure to evaluate an estimator, the closer its value to zero, the closer the estimator is close to the true parameter $\theta$.

### Statistical Bias

The MSE provides a natural metric, but we can easily imagine multiple different phenomena that might make it large. Two fundamentally important are fluctuation in the estimator due to randomness in the dataset, and systematic error in the estimator due to the estimation procedure.

First, let's measure the systematic error. For an estimator $\hat{\theta}_n$, the mathematical illustration of *statistical bias* can be defined as

$$\text{bias}(\hat{\theta}_n) = E(\hat{\theta}_n - \theta) = E(\hat{\theta}_n) - \theta. \tag{A.2}$$

Note that when $\text{bias}(\hat{\theta}_n) = 0$, the expectation of the estimator $\hat{\theta}_n$ is equal to the true value of parameter. In this case, we say $\hat{\theta}_n$ is an unbiased estimator. In general, an unbiased estimator is better than a biased estimator since its expectation is the same as the true parameter.

It is worth being aware, however, that biased estimators are frequently used in practice. There are cases where unbiased estimators do not exist without further assumptions, or are intractable to compute. This may seem like a significant flaw in an estimator, however the majority of estimators encountered in practice are at least asymptotically unbiased in the sense that the bias tends to zero as the number of available samples tends to infinity: $\lim_{n \rightarrow \infty} \text{bias}(\hat{\theta}_n) = 0$.

### Variance and Standard Deviation

Second, let's measure the randomness in the estimator. Recall from Section A.6, the *standard deviation* (or *standard error*) is defined as the squared root of the variance. We may measure the degree of fluctuation of an estimator by measuring the standard deviation or variance of that estimator.

$$\sigma_{\hat{\theta}_n} = \sqrt{\text{Var}(\hat{\theta}_n)} = \sqrt{E[(\hat{\theta}_n - E(\hat{\theta}_n))^2]}. \tag{A.3}$$

It is important to compare (A.3) to (A.1). In this equation we do not compare to the true population value $\theta$, but instead to $E(\hat{\theta}_n)$, the expected sample mean. Thus we are not measuring how far the estimator tends to be from the true value, but instead we measuring the fluctuation of the estimator itself.

### The Bias-Variance Trade-off

It is intuitively clear that these two main components contribute to the mean squared error. What is somewhat shocking is that we can show that this is actually a *decomposition* of the mean squared error into these two contributions plus a third one. That is to say that we can write the mean squared error as the sum of the square of the bias, the variance and the irreducible error.

$$\begin{aligned}
\mathrm{MSE}(\hat{\theta}_n, \theta) &= E[(\hat{\theta}_n - \theta)^2] \\
&= E[(\hat{\theta}_n)^2] + E[\theta^2] - 2E[\hat{\theta}_n\theta] \\
&= \mathrm{Var}[\hat{\theta}_n] + E[\hat{\theta}_n]^2 + \mathrm{Var}[\theta] + E[\theta]^2 - 2E[\hat{\theta}_n]E[\theta] \\
&= (E[\hat{\theta}_n] - E[\theta])^2 + \mathrm{Var}[\hat{\theta}_n] + \mathrm{Var}[\theta] \\
&= (E[\hat{\theta}_n - \theta])^2 + \mathrm{Var}[\hat{\theta}_n] + \mathrm{Var}[\theta] \\
&= (\mathrm{bias}[\hat{\theta}_n])^2 + \mathrm{Var}(\hat{\theta}_n) + \mathrm{Var}[\theta].
\end{aligned} \tag{A.4}$$

We refer the above formula as *bias-variance trade-off*. The mean squared error can be divided into three sources of error: the error from high bias, the error from high variance and the irreducible error. The bias error is commonly seen in a simple model (such as a linear regression model), which cannot extract high dimensional relations between the features and the outputs. If a model suffers from high bias error, we often say it is *underfitting* or lack of *flexibilty* as introduced in (Section 3.6). The high variance usually results from a too complex model, which overfits the training data. As a result, an *overfitting* model is sensitive to small fluctuations in the data. If a model suffers from high variance, we often say it is *overfitting* and lack of *generalization* as introduced in (Section 3.6). The irreducible error is the result from noise in the $\theta$ itself.

### Evaluating Estimators in Code

Since the standard deviation of an estimator has been implementing by simply calling `a.std()` for a tensor `a`, we will skip it but implement the statistical bias and the mean squared error.

```python
# Statistical bias
def stat_bias(true_theta, est_theta):
    return(torch.mean(est_theta) - true_theta)

# Mean squared error
def mse(data, true_theta):
    return(torch.mean(torch.square(data - true_theta)))
```

To illustrate the equation of the bias-variance trade-off, let's simulate of normal distribution $\mathcal{N}(\theta, \sigma^2)$ with $10,000$ samples. Here, we use a $\theta = 1$ and $\sigma = 4$. As the estimator is a function of the given samples, here we use the mean of the samples as an estimator for true $\theta$ in this normal distribution $\mathcal{N}(\theta, \sigma^2)$ .

```
theta_true = 1
sigma = 4
sample_len = 10000
samples = torch.normal(theta_true, sigma, size=(sample_len, 1))
theta_est = torch.mean(samples)
theta_est
```

```
tensor(1.0170)
```

Let's validate the trade-off equation by calculating the summation of the squared bias and the variance of our estimator. First, calculate the MSE of our estimator.

```
mse(samples, theta_true)
```

```
tensor(16.0298)
```

Next, we calculate $\mathrm{Var}(\hat{\theta}_n) + [\mathrm{bias}(\hat{\theta}_n)]^2$ as below. As you can see, the two values agree to numerical precision.

```
bias = stat_bias(theta_true, theta_est)
torch.square(samples.std(unbiased=False)) + torch.square(bias)
```

```
tensor(16.0298)
```

## A.10.2 Conducting Hypothesis Tests

The most commonly encountered topic in statistical inference is hypothesis testing. While hypothesis testing was popularized in the early 20th century, the first use can be traced back to John Arbuthnot in the 1700s. John tracked 80-year birth records in London and concluded that more men were born than women each year. Following that, the modern significance testing is the intelligence heritage by Karl Pearson who invented $p$-value and Pearson's chi-squared test, William Gosset who is the father of Student's t-distribution, and Ronald Fisher who initialed the null hypothesis and the significance test.

A *hypothesis test* is a way of evaluating some evidence against the default statement about a population. We refer the default statement as the *null hypothesis* $H_0$, which we try to reject using the observed data. Here, we use $H_0$ as a starting point for the statistical significance testing. The *alternative hypothesis* $H_A$ (or $H_1$) is a statement that is contrary to the null hypothesis. A null hypothesis is often stated in a declarative form which posits a relationship between variables. It should reflect the brief as explicit as possible, and be testable by statistics theory.

Imagine you are a chemist. After spending thousands of hours in the lab, you develop a new medicine which can dramatically improve one's ability to understand math. To show its magic power, you need to test it. Naturally, you may need some volunteers to take the medicine and see whether it can help them learn mathematics better. How do you get started?

First, you will need carefully random selected two groups of volunteers, so that there is no difference between their mathematical understanding ability measured by some metrics. The two groups are commonly referred to as the test group and the control group. The *test group* (or *treatment group*) is a group of individuals who will experience the medicine, while the *control group* represents the group of users who are set aside as a benchmark, i.e., identical environment setups except taking this medicine. In this way, the influence of all the variables are minimized, except the impact of the independent variable in the treatment.

Second, after a period of taking the medicine, you will need to measure the two groups' mathematical understanding by the same metrics, such as letting the volunteers do the same tests after learning a new mathematical formula. Then, you can collect their performance and compare the results. In this case, our null hypothesis will be that there is no difference between the two groups, and our alternate will be that there is.

This is still not fully formal. There are many details you have to think of carefully. For example, what is the suitable metrics to test their mathematical understanding ability? How many volunteers for your test so you can be confident to claim the effectiveness of your medicine? How long should you run the test? How do you decide if there is a difference between the two groups? Do you care about the average performance only, or also the range of variation of the scores? And so on.

In this way, hypothesis testing provides a framework for experimental design and reasoning about certainty in observed results. If we can now show that the null hypothesis is very unlikely to be true, we may reject it with confidence.

To complete the story of how to work with hypothesis testing, we need to now introduce some additional terminology and make some of our concepts above formal.

### Statistical Significance

The *statistical significance* measures the probability of erroneously rejecting the null hypothesis, $H_0$, when it should not be rejected, i.e.,

$$\text{statistical significance } = 1 - \alpha = 1 - P(\text{reject } H_0 \mid H_0 \text{ is true}). \tag{A.5}$$

It is also referred to as the *type I error* or *false positive*. The $\alpha$, is called as the *significance level* and its commonly used value is $5\%$, i.e., $1-\alpha = 95\%$. The significance level can be explained as the level of risk that we are willing to take, when we reject a true null hypothesis.

Fig. A.1 shows the observations' values and probability of a given normal distribution in a two-sample hypothesis test. If the observation data example is located outsides the 95% threshold, it will be a very unlikely observation under the null hypothesis assumption. Hence, there might be something wrong with the null hypothesis and we will reject it.

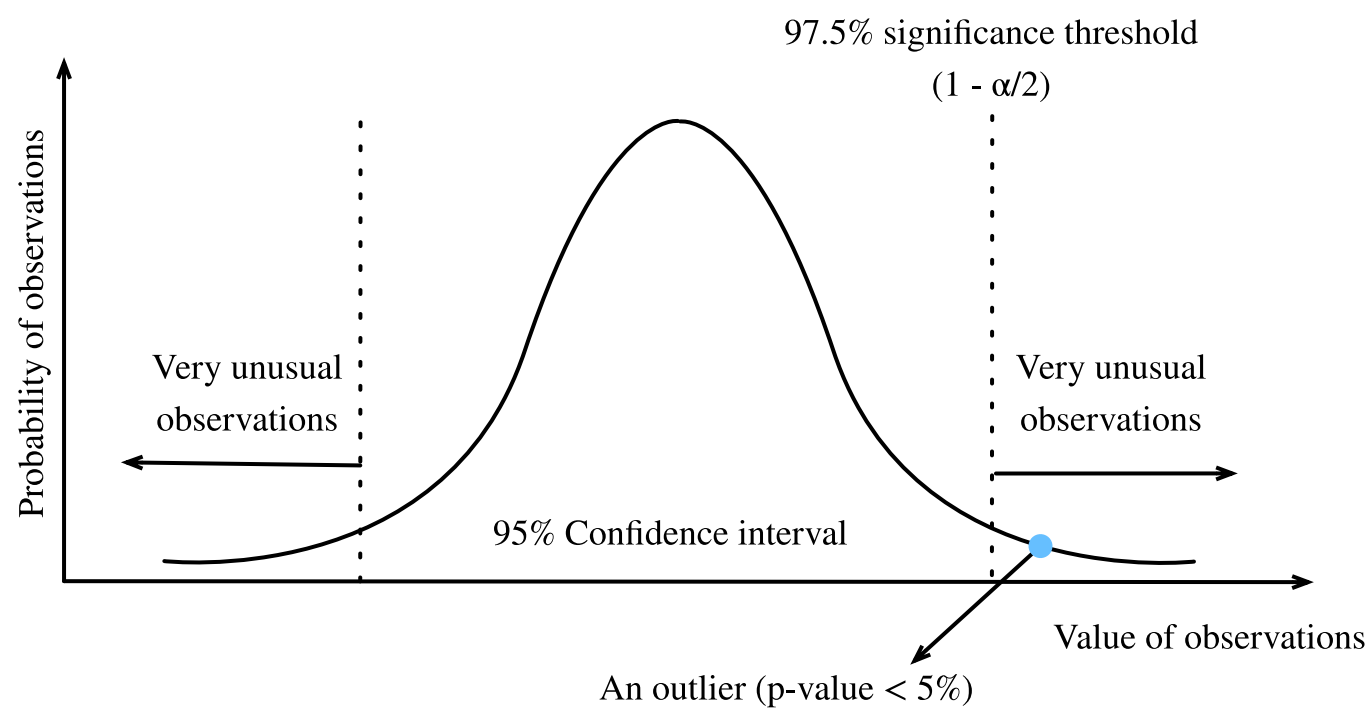

Fig. A.1 Statistical significance.

## Statistical Power

The *statistical power* (or *sensitivity*) measures the probability of reject the null hypothesis, $H_0$, when it should be rejected, i.e.,

$$\text{statistical power } = 1 - \beta = 1 - P(\text{ fail to reject } H_0 \mid H_0 \text{ is false}). \tag{A.6}$$

Recall that a *type I error* is error caused by rejecting the null hypothesis when it is true, whereas a *type II error* is resulted from failing to reject the null hypothesis when it is false. A type II error is usually denoted as $\beta$, and hence the corresponding statistical power is $1 - \beta$.

Intuitively, statistical power can be interpreted as how likely our test will detect a real discrepancy of some minimum magnitude at a desired statistical significance level. 80% is a commonly used statistical power threshold. The higher the statistical power, the more likely we are to detect true differences.

One of the most common uses of statistical power is in determining the number of samples needed. The probability you reject the null hypothesis when it is false depends on the degree to which it is false (known as the *effect size*) and the number of samples you have. As you might expect, small effect sizes will require a very large number of samples to be detectable with high probability. While beyond the scope of this brief appendix to derive in detail, as an example, want to be able to reject a null hypothesis that our sample came from a mean zero variance one Gaussian, and we believe that our sample's mean is actually close to one, we can do so with acceptable error rates with a sample size of only 8. However, if we think our sample population true mean is close to 0.01, then we'd need a sample size of nearly 80000 to detect the difference.

We can imagine the power as a water filter. In this analogy, a high power hypothesis test is like a high quality water filtration system that will reduce harmful substances in the water as much as possible. On the other hand, a smaller discrepancy is like a low quality water filter, where some relative small substances may easily escape from the gaps. Similarly, if the statistical power is not of enough high power, then the test may not catch the smaller discrepancy.

### Test Statistic

A *test statistic* $T(x)$ is a scalar which summarizes some characteristic of the sample data. The goal of defining such a statistic is that it should allow us to distinguish between different distributions and conduct our hypothesis test. Thinking back to our chemist example, if we wish to show that one population performs better than the other, it could be reasonable to take the mean as the test statistic. Different choices of test statistic can lead to statistical test with drastically different statistical power.

Often, $T(X)$ (the distribution of the test statistic under our null hypothesis) will follow, at least approximately, a common probability distribution such as a normal distribution when considered under the null hypothesis. If we can derive explicitly such a distribution, and then measure our test statistic on our dataset, we can safely reject the null hypothesis if our statistic is far outside the range that we would expect. Making this quantitative leads us to the notion of $p$-values.

### $p$-value

The $p$-value (or the *probability value*) is the probability that $T(X)$ is at least as extreme as the observed test statistic $T(x)$ assuming that the null hypothesis is *true*, i.e.,

$$p\text{-value} = P_{H_0}(T(X) \geq T(x)). \tag{A.7}$$

If the $p$-value is smaller than or equal to a predefined and fixed statistical significance level $\alpha$, we may reject the null hypothesis. Otherwise, we will conclude that we are lack of evidence to reject the null hypothesis. For a given population distribution, the *region of rejection* will be the interval contained of all the points which has a $p$-value smaller than the statistical significance level $\alpha$.

### One-side Test and Two-sided Test

Normally there are two kinds of significance test: the one-sided test and the two-sided test. The *one-sided test* (or *one-tailed test*) is applicable when the null hypothesis and the alternative hypothesis only have one direction. For example, the null hypothesis may state that the true parameter $\theta$ is less than or equal to a value $c$. The alternative hypothesis would be that $\theta$ is greater than $c$. That is, the region of rejection is on only one side of the sampling distribution. Contrary to the one-sided test, the *two-sided test* (or *two-tailed test*) is applicable when the region of rejection is on both sides of the sampling distribution. An example in this case may have a null hypothesis state that the true parameter $\theta$ is equal to a value $c$. The alternative hypothesis would be that $\theta$ is not equal to $c$.

### General Steps of Hypothesis Testing

After getting familiar with the above concepts, let's go through the general steps of hypothesis testing.

1. State the question and establish a null hypotheses $H_0$.
2. Set the statistical significance level $\alpha$ and a statistical power $(1 - \beta)$.
3. Obtain samples through experiments. The number of samples needed will depend on the statistical power, and the expected effect size.
4. Calculate the test statistic and the $p$-value.
5. Make the decision to keep or reject the null hypothesis based on the $p$-value and the statistical significance level $\alpha$.

To conduct a hypothesis test, we start by defining a null hypothesis and a level of risk that we are willing to take. Then we calculate the test statistic of the sample, taking an extreme value of the test statistic as evidence against the null hypothesis. If the test statistic falls within the reject region, we may reject the null hypothesis in favor of the alternative.

Hypothesis testing is applicable in a variety of scenarios such as the clinical trails and A/B testing.

## A.10.3 Constructing Confidence Intervals

When estimating the value of a parameter $\theta$, point estimators like $\hat{\theta}$ are of limited utility since they contain no notion of uncertainty. Rather, it would be far better if we could produce an interval that would contain the true parameter $\theta$ with high probability. If you were interested in such ideas a century ago, then you would have been excited to read "Outline of a Theory of Statistical Estimation Based on the Classical Theory of Probability" by Jerzy Neyman (Neyman, 1937), who first introduced the concept of confidence interval in 1937.

To be useful, a confidence interval should be as small as possible for a given degree of certainty. Let's see how to derive it.

### Definition

Mathematically, a *confidence interval* for the true parameter $\theta$ is an interval $C_n$ that computed from the sample data such that

$$P_{\theta}(C_n \ni \theta) \geq 1 - \alpha, \forall \theta. \tag{A.8}$$

Here $\alpha \in (0, 1)$, and $1 - \alpha$ is called the *confidence level* or *coverage* of the interval. This is the same $\alpha$ as the significance level as we discussed about above.

Note that (A.8) is about variable $C_n$, not about the fixed $\theta$. To emphasize this, we write $P_\theta(C_n \ni \theta)$ rather than $P_\theta(\theta \in C_n)$.

## Interpretation

It is very tempting to interpret a $95\%$ confidence interval as an interval where you can be $95\%$ sure the true parameter lies, however this is sadly not true. The true parameter is fixed, and it is the interval that is random. Thus a better interpretation would be to say that if you generated a large number of confidence intervals by this procedure, $95\%$ of the generated intervals would contain the true parameter.

This may seem pedantic, but it can have real implications for the interpretation of the results. In particular, we may satisfy (A.8) by constructing intervals that we are *almost certain* do not contain the true value, as long as we only do so rarely enough. We close this section by providing three tempting but false statements. An in-depth discussion of these points can be found in Morey *et al.* (2016).

- **Fallacy 1**. Narrow confidence intervals mean we can estimate the parameter precisely.
- **Fallacy 2**. The values inside the confidence interval are more likely to be the true value than those outside the interval.
- **Fallacy 3**. The probability that a particular observed $95\%$ confidence interval contains the true value is $95\%$.

Sufficed to say, confidence intervals are subtle objects. However, if you keep the interpretation clear, they can be powerful tools.

## A Gaussian Example

Let's discuss the most classical example, the confidence interval for the mean of a Gaussian of unknown mean and variance. Suppose we collect $n$ samples $\{x_i\}_{i=1}^n$ from our Gaussian $\mathcal{N}(\mu, \sigma^2)$. We can compute estimators for the mean and variance by taking

$$\hat{\mu}_n = \frac{1}{n}\sum_{i=1}^n x_i \text{ and } \hat{\sigma}_n^2 = \frac{1}{n-1}\sum_{i=1}^n (x_i - \hat{\mu})^2. \tag{A.9}$$

If we now consider the random variable

$$T = \frac{\hat{\mu}_n - \mu}{\hat{\sigma}_n/\sqrt{n}}, \tag{A.10}$$

we obtain a random variable following a well-known distribution called the *Student's t-distribution on* $n-1$ *degrees of freedom*.

This distribution is very well studied, and it is known, for instance, that as $n \to \infty$, it is approximately a standard Gaussian, and thus by looking up values of the Gaussian c.d.f. in a table, we may conclude that the value of $T$ is in the interval $[-1.96, 1.96]$ at least $95\%$ of the time. For finite values of $n$, the interval needs to be somewhat larger, but are well known and precomputed in tables.

Thus, we may conclude that for large $n$,

$$P\left(\frac{\hat{\mu}_n - \mu}{\hat{\sigma}_n/\sqrt{n}} \in [-1.96, 1.96]\right) \geq 0.95. \tag{A.11}$$

Rearranging this by multiplying both sides by $\hat{\sigma}_n/\sqrt{n}$ and then adding $\hat{\mu}_n$, we obtain

$$P\left(\mu \in \left[\hat{\mu}_n - 1.96\frac{\hat{\sigma}_n}{\sqrt{n}}, \hat{\mu}_n + 1.96\frac{\hat{\sigma}_n}{\sqrt{n}}\right]\right) \geq 0.95. \tag{A.12}$$

Thus we know that we have found our $95\%$ confidence interval:

$$\left[\hat{\mu}_n - 1.96\frac{\hat{\sigma}_n}{\sqrt{n}}, \hat{\mu}_n + 1.96\frac{\hat{\sigma}_n}{\sqrt{n}}\right]. \tag{A.13}$$

It is safe to say that (A.13) is one of the most used formula in statistics. Let's close our discussion of statistics by implementing it. For simplicity, we assume we are in the asymptotic regime. Small values of $N$ should include the correct value of `t_star` obtained either programmatically or from a $t$-table.

```python
# PyTorch uses Bessel's correction by default, which means the use of ddof=1
# instead of default ddof=0 in numpy. We can use unbiased=False to imitate
# ddof=0.

# Number of samples
N = 1000

# Sample dataset
samples = torch.normal(0, 1, size=(N,))

# Lookup Students's t-distribution c.d.f.
t_star = 1.96

# Construct interval
mu_hat = torch.mean(samples)
sigma_hat = samples.std(unbiased=True)
(mu_hat - t_star*sigma_hat/torch.sqrt(torch.tensor(N, dtype=torch.float32)),\
 mu_hat + t_star*sigma_hat/torch.sqrt(torch.tensor(N, dtype=torch.float32)))
```

```
(tensor(-0.0568), tensor(0.0704))
```

## A.10.4 Summary

- Statistics focuses on inference problems, whereas deep learning emphasizes on making accurate predictions without explicitly programming and understanding.

- There are three common statistics inference methods: evaluating and comparing estimators, conducting hypothesis tests, and constructing confidence intervals.
- There are three most common estimators: statistical bias, standard deviation, and mean square error.
- A confidence interval is an estimated range of a true population parameter that we can construct by given the samples.
- Hypothesis testing is a way of evaluating some evidence against the default statement about a population.

### A.10.5 Exercises

1. Let $X_1, X_2, \ldots, X_n \stackrel{\text{iid}}{\sim} \mathrm{Unif}(0, \theta)$, where "iid" stands for *independent and identically distributed*. Consider the following estimators of $\theta$:

$$\hat{\theta} = \max\{X_1, X_2, \ldots, X_n\}; \tag{A.14}$$

$$\tilde{\theta} = 2\bar{X}_n = \frac{2}{n}\sum_{i=1}^{n} X_i. \tag{A.15}$$

   - Find the statistical bias, standard deviation, and mean square error of $\hat{\theta}$.
   - Find the statistical bias, standard deviation, and mean square error of $\tilde{\theta}$.
   - Which estimator is better?

2. For our chemist example in introduction, can you derive the 5 steps to conduct a two-sided hypothesis testing? Given the statistical significance level $\alpha = 0.05$ and the statistical power $1 - \beta = 0.8$.
3. Run the confidence interval code with $N = 2$ and $\alpha = 0.5$ for 100 independently generated dataset, and plot the resulting intervals (in this case `t_star = 1.0`). You will see several very short intervals which are very far from containing the true mean 0. Does this contradict the interpretation of the confidence interval? Do you feel comfortable using short intervals to indicate high precision estimates?

Discussions[287].

## A.11 Information Theory

The universe is overflowing with information. Information provides a common language across disciplinary rifts: from Shakespeare's Sonnet to researchers' paper on Cornell ArXiv, from Van Gogh's printing Starry Night to Beethoven's music Symphony No. 5, from the first programming language Plankalkül to the state-of-the-art machine learning algorithms. Everything must follow the rules of information theory, no matter the format. With information theory, we can measure and compare how much information is present in different signals. In this section, we will investigate the fundamental concepts of information theory and applications of information theory in machine learning.

Before we get started, let's outline the relationship between machine learning and information theory. Machine learning aims to extract interesting signals from data and make critical predictions. On the other hand, information theory studies encoding, decoding, transmitting, and manipulating information. As a result, information theory provides fundamental language for discussing the information processing in machine learned systems. For example, many machine learning applications use the cross-entropy loss as described in Section 4.1. This loss can be directly derived from information theoretic considerations.

## A.11.1 Information

Let's start with the "soul" of information theory: information. *Information* can be encoded in anything with a particular sequence of one or more encoding formats. Suppose that we task ourselves with trying to define a notion of information. What could be our starting point?

Consider the following thought experiment. We have a friend with a deck of cards. They will shuffle the deck, flip over some cards, and tell us statements about the cards. We will try to assess the information content of each statement.

First, they flip over a card and tell us, "I see a card." This provides us with no information at all. We were already certain that this was the case so we hope the information should be zero.

Next, they flip over a card and say, "I see a heart." This provides us some information, but in reality there are only $4$ different suits that were possible, each equally likely, so we are not surprised by this outcome. We hope that whatever the measure of information, this event should have low information content.

Next, they flip over a card and say, "This is the $3$ of spades." This is more information. Indeed there were $52$ equally likely possible outcomes, and our friend told us which one it was. This should be a medium amount of information.

Let's take this to the logical extreme. Suppose that finally they flip over every card from the deck and read off the entire sequence of the shuffled deck. There are $52!$ different orders to the deck, again all equally likely, so we need a lot of information to know which one it is.

Any notion of information we develop must conform to this intuition. Indeed, in the next sections we will learn how to compute that these events have $0$ bits, $2$ bits, $~5.7$ bits, and $~225.6$ bits of information respectively.

If we read through these thought experiments, we see a natural idea. As a starting point, rather than caring about the knowledge, we may build off the idea that information represents the degree of surprise or the abstract possibility of the event. For example, if we want to describe an unusual event, we need a lot information. For a common event, we may not need much information.

In 1948, Claude E. Shannon published *A Mathematical Theory of Communication* (Shannon, 1948) establishing the theory of information. In his article, Shannon introduced the concept of information entropy for the first time. We will begin our journey here.

## Self-information

Since information embodies the abstract possibility of an event, how do we map the possibility to the number of bits? Shannon introduced the terminology *bit* as the unit of information, which was originally created by John Tukey. So what is a "bit" and why do we use it to measure information? Historically, an antique transmitter can only send or receive two types of code: $0$ and $1$. Indeed, binary encoding is still in common use on all modern digital computers. In this way, any information is encoded by a series of $0$ and $1$. And hence, a series of binary digits of length $n$ contains $n$ bits of information.

Now, suppose that for any series of codes, each $0$ or $1$ occurs with a probability of $\frac{1}{2}$. Hence, an event $X$ with a series of codes of length $n$, occurs with a probability of $\frac{1}{2^n}$. At the same time, as we mentioned before, this series contains $n$ bits of information. So, can we generalize to a mathematical function which can transfer the probability $p$ to the number of bits? Shannon gave the answer by defining *self-information*

$$I(X) = -\log_2(p), \tag{A.1}$$

as the *bits* of information we have received for this event $X$. Note that we will always use base-2 logarithms in this section. For the sake of simplicity, the rest of this section will omit the subscript 2 in the logarithm notation, i.e., $\log(.)$ always refers to $\log_2(.)$. For example, the code "0010" has a self-information

$$I(\text{"0010"}) = -\log(p(\text{"0010"})) = -\log\left(\frac{1}{2^4}\right) = 4 \text{ bits}. \tag{A.2}$$

We can calculate self information as shown below. Before that, let's first import all the necessary packages in this section.

```
import torch
from torch.nn import NLLLoss

def nansum(x):
    # Define nansum, as pytorch does not offer it inbuilt.
    return x[~torch.isnan(x)].sum()

def self_information(p):
    return -torch.log2(torch.tensor(p)).item()
```

(continued from previous page)

```
self_information(1 / 64)
```

```
6.0
```

## A.11.2 Entropy

As self-information only measures the information of a single discrete event, we need a more generalized measure for any random variable of either discrete or continuous distribution.

### Motivating Entropy

Let's try to get specific about what we want. This will be an informal statement of what are known as the *axioms of Shannon entropy*. It will turn out that the following collection of common-sense statements force us to a unique definition of information. A formal version of these axioms, along with several others may be found in Csiszár (2008).

1. The information we gain by observing a random variable does not depend on what we call the elements, or the presence of additional elements which have probability zero.
2. The information we gain by observing two random variables is no more than the sum of the information we gain by observing them separately. If they are independent, then it is exactly the sum.
3. The information gained when observing (nearly) certain events is (nearly) zero.

While proving this fact is beyond the scope of our text, it is important to know that this uniquely determines the form that entropy must take. The only ambiguity that these allow is in the choice of fundamental units, which is most often normalized by making the choice we saw before that the information provided by a single fair coin flip is one bit.

### Definition

For any random variable $X$ that follows a probability distribution $P$ with a probability density function (p.d.f.) or a probability mass function (p.m.f.) $p(x)$, we measure the expected amount of information through *entropy* (or *Shannon entropy*)

$$H(X) = -E_{x \sim P} [\log p(x)]. \tag{A.3}$$

To be specific, if $X$ is discrete,

$$H(X) = - \sum_i p_i \log p_i, \text{ where } p_i = P(X_i). \tag{A.4}$$

Otherwise, if $X$ is continuous, we also refer entropy as *differential entropy*

$$H(X) = -\int_x p(x) \log p(x) \; dx. \tag{A.5}$$

We can define entropy as below.

```
def entropy(p):
    entropy = - p * torch.log2(p)
    # Operator `nansum` will sum up the non-nan number
    out = nansum(entropy)
    return out

entropy(torch.tensor([0.1, 0.5, 0.1, 0.3]))
```

```
tensor(1.6855)
```

## Interpretations

You may be curious: in the entropy definition (A.3), why do we use an expectation of a negative logarithm? Here are some intuitions.

First, why do we use a *logarithm* function log? Suppose that $p(x) = f_1(x) f_2(x) \ldots, f_n(x)$, where each component function $f_i(x)$ is independent from each other. This means that each $f_i(x)$ contributes independently to the total information obtained from $p(x)$. As discussed above, we want the entropy formula to be additive over independent random variables. Luckily, log can naturally turn a product of probability distributions to a summation of the individual terms.

Next, why do we use a *negative* log? Intuitively, more frequent events should contain less information than less common events, since we often gain more information from an unusual case than from an ordinary one. However, log is monotonically increasing with the probabilities, and indeed negative for all values in $[0, 1]$. We need to construct a monotonically decreasing relationship between the probability of events and their entropy, which will ideally be always positive (for nothing we observe should force us to forget what we have known). Hence, we add a negative sign in front of log function.

Last, where does the *expectation* function come from? Consider a random variable $X$. We can interpret the self-information ($-\log(p)$) as the amount of *surprise* we have at seeing a particular outcome. Indeed, as the probability approaches zero, the surprise becomes infinite. Similarly, we can interpret the entropy as the average amount of surprise from observing $X$. For example, imagine that a slot machine system emits statistical independently symbols $s_1, \ldots, s_k$ with probabilities $p_1, \ldots, p_k$ respectively. Then the entropy of this system equals to the average self-information from observing each output, i.e.,

$$H(S) = \sum_i p_i \cdot I(s_i) = - \sum_i p_i \cdot \log p_i. \tag{A.6}$$

### Properties of Entropy

By the above examples and interpretations, we can derive the following properties of entropy (A.3). Here, we refer to $X$ as an event and $P$ as the probability distribution of $X$.

- $H(X) \geq 0$ for all discrete $X$ (entropy can be negative for continuous $X$).
- If $X \sim P$ with a p.d.f. or a p.m.f. $p(x)$, and we try to estimate $P$ by a new probability distribution $Q$ with a p.d.f. or a p.m.f. $q(x)$, then

$$H(X) = -E_{x \sim P}[\log p(x)] \leq -E_{x \sim P}[\log q(x)], \text{ with equality if and only if } P = Q. \tag{A.7}$$

  Alternatively, $H(X)$ gives a lower bound of the average number of bits needed to encode symbols drawn from $P$.

- If $X \sim P$, then $x$ conveys the maximum amount of information if it spreads evenly among all possible outcomes. Specifically, if the probability distribution $P$ is discrete with $k$-class $\{p_1, \ldots, p_k\}$, then

$$H(X) \leq \log(k), \text{ with equality if and only if } p_i = \frac{1}{k}, \forall i. \tag{A.8}$$

  If $P$ is a continuous random variable, then the story becomes much more complicated. However, if we additionally impose that $P$ is supported on a finite interval (with all values between 0 and 1), then $P$ has the highest entropy if it is the uniform distribution on that interval.

## A.11.3 Mutual Information

Previously we defined entropy of a single random variable $X$, how about the entropy of a pair random variables $(X, Y)$? We can think of these techniques as trying to answer the following type of question, "What information is contained in $X$ and $Y$ together compared to each separately? Is there redundant information, or is it all unique?"

For the following discussion, we always use $(X, Y)$ as a pair of random variables that follows a joint probability distribution $P$ with a p.d.f. or a p.m.f. $p_{X,Y}(x, y)$, while $X$ and $Y$ follow probability distribution $p_X(x)$ and $p_Y(y)$, respectively.

### Joint Entropy

Similar to entropy of a single random variable (A.3), we define the *joint entropy* $H(X, Y)$ of a pair random variables $(X, Y)$ as

$$H(X, Y) = -E_{(x,y) \sim P}[\log p_{X,Y}(x, y)]. \tag{A.9}$$

Precisely, on the one hand, if $(X, Y)$ is a pair of discrete random variables, then

$$H(X, Y) = -\sum_x \sum_y p_{X,Y}(x, y) \log p_{X,Y}(x, y). \tag{A.10}$$

On the other hand, if $(X, Y)$ is a pair of continuous random variables, then we define the *differential joint entropy* as

$$H(X, Y) = -\int_{x,y} p_{X,Y}(x, y) \ \log p_{X,Y}(x, y) \ dx \ dy. \tag{A.11}$$

We can think of (A.9) as telling us the total randomness in the pair of random variables. As a pair of extremes, if $X = Y$ are two identical random variables, then the information in the pair is exactly the information in one and we have $H(X, Y) = H(X) = H(Y)$. On the other extreme, if $X$ and $Y$ are independent then $H(X, Y) = H(X) + H(Y)$. Indeed we will always have that the information contained in a pair of random variables is no smaller than the entropy of either random variable and no more than the sum of both.

$$H(X), H(Y) \le H(X, Y) \le H(X) + H(Y). \tag{A.12}$$

Let's implement joint entropy from scratch.

```
def joint_entropy(p_xy):
    joint_ent = -p_xy * torch.log2(p_xy)
    # Operator `nansum` will sum up the non-nan number
    out = nansum(joint_ent)
    return out

joint_entropy(torch.tensor([[0.1, 0.5], [0.1, 0.3]]))
```

```
tensor(1.6855)
```

Notice that this is the same *code* as before, but now we interpret it differently as working on the joint distribution of the two random variables.

## Conditional Entropy

The joint entropy defined above the amount of information contained in a pair of random variables. This is useful, but oftentimes it is not what we care about. Consider the setting of machine learning. Let's take $X$ to be the random variable (or vector of random variables) that describes the pixel values of an image, and $Y$ to be the random variable which is the class label. $X$ should contain substantial information—a natural image is a complex thing. However, the information contained in $Y$ once the image has been show should be low. Indeed, the image of a digit should already contain the information about what digit it is unless the digit is illegible. Thus, to continue to extend our vocabulary of information theory, we need to be able to reason about the information content in a random variable conditional on another.

In the probability theory, we saw the definition of the *conditional probability* to measure the relationship between variables. We now want to analogously define the *conditional entropy* $H(Y \mid X)$. We can write this as

$$H(Y \mid X) = -E_{(x,y)\sim P}[\log p(y \mid x)], \tag{A.13}$$

where $p(y \mid x) = \frac{p_{X,Y}(x,y)}{p_X(x)}$ is the conditional probability. Specifically, if $(X, Y)$ is a pair of discrete random variables, then

$$H(Y \mid X) = - \sum_x \sum_y p(x, y) \log p(y \mid x). \tag{A.14}$$

If $(X, Y)$ is a pair of continuous random variables, then the *differential conditional entropy* is similarly defined as

$$H(Y \mid X) = - \int_x \int_y p(x, y) \ \log p(y \mid x) \ dx \ dy. \tag{A.15}$$

It is now natural to ask, how does the *conditional entropy* $H(Y \mid X)$ relate to the entropy $H(X)$ and the joint entropy $H(X, Y)$? Using the definitions above, we can express this cleanly:

$$H(Y \mid X) = H(X, Y) - H(X). \tag{A.16}$$

This has an intuitive interpretation: the information in $Y$ given $X$ ($H(Y \mid X)$) is the same as the information in both $X$ and $Y$ together ($H(X, Y)$) minus the information already contained in $X$. This gives us the information in $Y$ which is not also represented in $X$.

Now, let's implement conditional entropy (A.13) from scratch.

```
def conditional_entropy(p_xy, p_x):
    p_y_given_x = p_xy/p_x
    cond_ent = -p_xy * torch.log2(p_y_given_x)
    # Operator `nansum` will sum up the non-nan number
    out = nansum(cond_ent)
    return out

conditional_entropy(torch.tensor([[0.1, 0.5], [0.2, 0.3]]),
                    torch.tensor([0.2, 0.8]))
```

```
tensor(0.8635)
```

## Mutual Information

Given the previous setting of random variables $(X, Y)$, you may wonder: "Now that we know how much information is contained in $Y$ but not in $X$, can we similarly ask how much information is shared between $X$ and $Y$?" The answer will be the *mutual information* of $(X, Y)$, which we will write as $I(X, Y)$.

Rather than diving straight into the formal definition, let's practice our intuition by first trying to derive an expression for the mutual information entirely based on terms we have constructed before. We wish to find the information shared between two random variables. One way we could try to do this is to start with all the information contained in both $X$ and $Y$ together, and then we take off the parts that are not shared. The information contained in both $X$ and $Y$ together is written as $H(X, Y)$. We want to subtract from this the information contained in $X$ but not in $Y$, and the information contained in $Y$ but not in $X$. As we saw in the previous section, this is given by $H(X \mid Y)$ and $H(Y \mid X)$ respectively. Thus, we have that the mutual information should be

$$I(X,Y) = H(X,Y) - H(Y \mid X) - H(X \mid Y). \tag{A.17}$$

Indeed, this is a valid definition for the mutual information. If we expand out the definitions of these terms and combine them, a little algebra shows that this is the same as

$$I(X,Y) = E_x E_y \left\{ p_{X,Y}(x,y) \log \frac{p_{X,Y}(x,y)}{p_X(x) p_Y(y)} \right\}. \tag{A.18}$$

We can summarize all of these relationships in image Fig. A.1. It is an excellent test of intuition to see why the following statements are all also equivalent to $I(X,Y)$.

- $H(X) - H(X \mid Y)$
- $H(Y) - H(Y \mid X)$
- $H(X) + H(Y) - H(X,Y)$

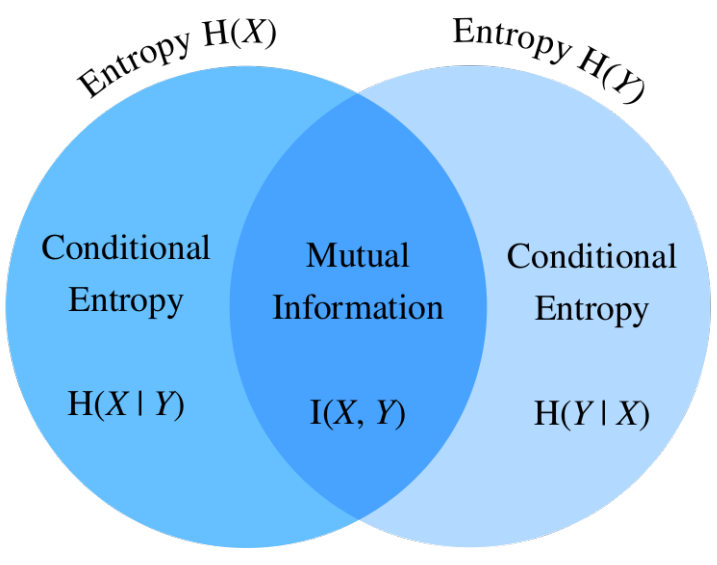

Fig. A.1 Mutual information's relationship with joint entropy and conditional entropy.

In many ways we can think of the mutual information (A.18) as principled extension of correlation coefficient we saw in Section A.6. This allows us to ask not only for linear relationships between variables, but for the maximum information shared between the two random variables of any kind.

Now, let's implement mutual information from scratch.

```
def mutual_information(p_xy, p_x, p_y):
    p = p_xy / (p_x * p_y)
    mutual = p_xy * torch.log2(p)
    # Operator `nansum` will sum up the non-nan number
    out = nansum(mutual)
    return out

mutual_information(torch.tensor([[0.1, 0.5], [0.1, 0.3]]),
                   torch.tensor([0.2, 0.8]), torch.tensor([[0.75, 0.25]]))
```

```
tensor(0.7195)
```

### Properties of Mutual Information

Rather than memorizing the definition of mutual information (A.18), you only need to keep in mind its notable properties:

- Mutual information is symmetric, i.e., $I(X, Y) = I(Y, X)$.
- Mutual information is non-negative, i.e., $I(X, Y) \geq 0$.
- $I(X, Y) = 0$ if and only if $X$ and $Y$ are independent. For example, if $X$ and $Y$ are independent, then knowing $Y$ does not give any information about $X$ and vice versa, so their mutual information is zero.
- Alternatively, if $X$ is an invertible function of $Y$, then $Y$ and $X$ share all information and

$$I(X, Y) = H(Y) = H(X). \tag{A.19}$$

### Pointwise Mutual Information

When we worked with entropy at the beginning of this chapter, we were able to provide an interpretation of $-\log(p_X(x))$ as how *surprised* we were with the particular outcome. We may give a similar interpretation to the logarithmic term in the mutual information, which is often referred to as the *pointwise mutual information*:

$$\mathrm{pmi}(x, y) = \log \frac{p_{X,Y}(x, y)}{p_X(x) p_Y(y)}. \tag{A.20}$$

We can think of (A.20) as measuring how much more or less likely the specific combination of outcomes $x$ and $y$ are compared to what we would expect for independent random outcomes. If it is large and positive, then these two specific outcomes occur much more frequently than they would compared to random chance (*note*: the denominator is $p_X(x) p_Y(y)$ which is the probability of the two outcomes were independent), whereas if it is large and negative it represents the two outcomes happening far less than we would expect by random chance.

This allows us to interpret the mutual information (A.18) as the average amount that we were surprised to see two outcomes occurring together compared to what we would expect if they were independent.

### Applications of Mutual Information

Mutual information may be a little abstract in it pure definition, so how does it related to machine learning? In natural language processing, one of the most difficult problems is the *ambiguity resolution*, or the issue of the meaning of a word being unclear from context. For example, recently a headline in the news reported that "Amazon is on fire". You may wonder whether the company Amazon has a building on fire, or the Amazon rain forest is on fire.

In this case, mutual information can help us resolve this ambiguity. We first find the group of words that each has a relatively large mutual information with the company Amazon, such as e-commerce, technology, and online. Second, we find another group of words that each has a relatively large mutual information with the Amazon rain forest, such as rain, forest, and tropical. When we need to disambiguate "Amazon", we can compare which group has more occurrence in the context of the word Amazon. In this case the article would go on to describe the forest, and make the context clear.

## A.11.4 Kullback–Leibler Divergence

As what we have discussed in Section 2.3, we can use norms to measure distance between two points in space of any dimensionality. We would like to be able to do a similar task with probability distributions. There are many ways to go about this, but information theory provides one of the nicest. We now explore the *Kullback–Leibler (KL) divergence*, which provides a way to measure if two distributions are close together or not.

### Definition

Given a random variable $X$ that follows the probability distribution $P$ with a p.d.f. or a p.m.f. $p(x)$, and we estimate $P$ by another probability distribution $Q$ with a p.d.f. or a p.m.f. $q(x)$. Then the *Kullback–Leibler (KL) divergence* (or *relative entropy*) between $P$ and $Q$ is

$$D_{\mathrm{KL}}(P\|Q) = E_{x\sim P}\left[\log \frac{p(x)}{q(x)}\right]. \tag{A.21}$$

As with the pointwise mutual information (A.20), we can again provide an interpretation of the logarithmic term: $-\log \frac{q(x)}{p(x)} = -\log(q(x)) - (-\log(p(x)))$ will be large and positive if we see $x$ far more often under $P$ than we would expect for $Q$, and large and negative if we see the outcome far less than expected. In this way, we can interpret it as our *relative* surprise at observing the outcome compared to how surprised we would be observing it from our reference distribution.

Let's implement the KL divergence from Scratch.

```
def kl_divergence(p, q):
    kl = p * torch.log2(p / q)
    out = nansum(kl)
    return out.abs().item()
```

### KL Divergence Properties

Let's take a look at some properties of the KL divergence (A.21).

- KL divergence is non-symmetric, i.e., there are $P, Q$ such that

$$D_{\mathrm{KL}}(P\|Q) \neq D_{\mathrm{KL}}(Q\|P). \tag{A.22}$$

- KL divergence is non-negative, i.e.,

$$D_{\mathrm{KL}}(P\|Q) \geq 0. \tag{A.23}$$

  Note that the equality holds only when $P = Q$.

- If there exists an $x$ such that $p(x) > 0$ and $q(x) = 0$, then $D_{\mathrm{KL}}(P\|Q) = \infty$.
- There is a close relationship between KL divergence and mutual information. Besides the relationship shown in Fig. A.1, $I(X, Y)$ is also numerically equivalent with the following terms:
  1. $D_{\mathrm{KL}}(P(X, Y) \mid\mid P(X)P(Y))$;
  2. $E_Y\{D_{\mathrm{KL}}(P(X \mid Y) \mid\mid P(X))\}$;
  3. $E_X\{D_{\mathrm{KL}}(P(Y \mid X) \mid\mid P(Y))\}$.

For the first term, we interpret mutual information as the KL divergence between $P(X, Y)$ and the product of $P(X)$ and $P(Y)$, and thus is a measure of how different the joint distribution is from the distribution if they were independent. For the second term, mutual information tells us the average reduction in uncertainty about $Y$ that results from learning the value of the $X$'s distribution. Similarly to the third term.

### Example

Let's go through a toy example to see the non-symmetry explicitly.

First, let's generate and sort three tensors of length $10,000$: an objective tensor $p$ which follows a normal distribution $N(0, 1)$, and two candidate tensors $q_1$ and $q_2$ which follow normal distributions $N(-1, 1)$ and $N(1, 1)$ respectively.

```
torch.manual_seed(1)

tensor_len = 10000
p = torch.normal(0, 1, (tensor_len, ))
q1 = torch.normal(-1, 1, (tensor_len, ))
q2 = torch.normal(1, 1, (tensor_len, ))

p = torch.sort(p)[0]
q1 = torch.sort(q1)[0]
q2 = torch.sort(q2)[0]
```

Since $q_1$ and $q_2$ are symmetric with respect to the y-axis (i.e., $x = 0$), we expect a similar value of KL divergence between $D_{\mathrm{KL}}(p\|q_1)$ and $D_{\mathrm{KL}}(p\|q_2)$. As you can see below, there is only a less than 3% off between $D_{\mathrm{KL}}(p\|q_1)$ and $D_{\mathrm{KL}}(p\|q_2)$.

```
kl_pq1 = kl_divergence(p, q1)
kl_pq2 = kl_divergence(p, q2)
similar_percentage = abs(kl_pq1 - kl_pq2) / ((kl_pq1 + kl_pq2) / 2) * 100

kl_pq1, kl_pq2, similar_percentage
```

```
(8582.0341796875, 8828.3095703125, 2.8290698237936858)
```

In contrast, you may find that $D_{\mathrm{KL}}(q_2\|p)$ and $D_{\mathrm{KL}}(p\|q_2)$ are off a lot, with around 40% off as shown below.

```
kl_q2p = kl_divergence(q2, p)
differ_percentage = abs(kl_q2p - kl_pq2) / ((kl_q2p + kl_pq2) / 2) * 100

kl_q2p, differ_percentage
```

```
(14130.125, 46.18621024399691)
```

## A.11.5 Cross-Entropy

If you are curious about applications of information theory in deep learning, here is a quick example. We define the true distribution $P$ with probability distribution $p(x)$, and the estimated distribution $Q$ with probability distribution $q(x)$, and we will use them in the rest of this section.

Say we need to solve a binary classification problem based on given $n$ data examples $\{x_1, \ldots, x_n\}$. Assume that we encode 1 and 0 as the positive and negative class label $y_i$ respectively, and our neural network is parametrized by $\theta$. If we aim to find a best $\theta$ so that $\hat{y}_i = p_\theta(y_i \mid x_i)$, it is natural to apply the maximum log-likelihood approach as was seen in Section A.7. To be specific, for true labels $y_i$ and predictions $\hat{y}_i = p_\theta(y_i \mid x_i)$, the probability to be classified as positive is $\pi_i = p_\theta(y_i = 1 \mid x_i)$. Hence, the log-likelihood function would be

$$
\begin{aligned}
l(\theta) &= \log L(\theta) \\
&= \log \prod_{i=1}^{n} \pi_i^{y_i} (1 - \pi_i)^{1 - y_i} \\
&= \sum_{i=1}^{n} y_i \log(\pi_i) + (1 - y_i) \log(1 - \pi_i).
\end{aligned}
\tag{A.24}
$$

Maximizing the log-likelihood function $l(\theta)$ is identical to minimizing $-l(\theta)$, and hence we can find the best $\theta$ from here. To generalize the above loss to any distributions, we also called $-l(\theta)$ the *cross-entropy loss* $\mathrm{CE}(y, \hat{y})$, where $y$ follows the true distribution $P$ and $\hat{y}$ follows the estimated distribution $Q$.

This was all derived by working from the maximum likelihood point of view. However, if we look closely we can see that terms like $\log(\pi_i)$ have entered into our computation which is a solid indication that we can understand the expression from an information theoretic point of view.

### Formal Definition

Like KL divergence, for a random variable $X$, we can also measure the divergence between the estimating distribution $Q$ and the true distribution $P$ via *cross-entropy*,

$$\mathrm{CE}(P, Q) = -E_{x \sim P}[\log(q(x))]. \tag{A.25}$$

By using properties of entropy discussed above, we can also interpret it as the summation of the entropy $H(P)$ and the KL divergence between $P$ and $Q$, i.e.,

$$\mathrm{CE}(P, Q) = H(P) + D_{\mathrm{KL}}(P \| Q). \tag{A.26}$$

We can implement the cross-entropy loss as below.

```
def cross_entropy(y_hat, y):
    ce = -torch.log(y_hat[range(len(y_hat)), y])
    return ce.mean()
```

Now define two tensors for the labels and predictions, and calculate the cross-entropy loss of them.

```
labels = torch.tensor([0, 2])
preds = torch.tensor([[0.3, 0.6, 0.1], [0.2, 0.3, 0.5]])

cross_entropy(preds, labels)
```

```
tensor(0.9486)
```

### Properties

As alluded in the beginning of this section, cross-entropy (A.25) can be used to define a loss function in the optimization problem. It turns out that the following are equivalent:

1. Maximizing predictive probability of $Q$ for distribution $P$, (i.e., $E_{x \sim P}[\log(q(x))]$);
2. Minimizing cross-entropy $\mathrm{CE}(P, Q)$;
3. Minimizing the KL divergence $D_{\mathrm{KL}}(P \| Q)$.

The definition of cross-entropy indirectly proves the equivalent relationship between objective 2 and objective 3, as long as the entropy of true data $H(P)$ is constant.

### Cross-Entropy as An Objective Function of Multi-class Classification

If we dive deep into the classification objective function with cross-entropy loss CE, we will find minimizing CE is equivalent to maximizing the log-likelihood function $L$.

To begin with, suppose that we are given a dataset with $n$ examples, and it can be classified into $k$-classes. For each data example $i$, we represent any $k$-class label $\mathbf{y}_i = (y_{i1}, \ldots, y_{ik})$ by *one-hot encoding*. To be specific, if the example $i$ belongs to class $j$, then we set the $j$-th entry to 1, and all other components to 0, i.e.,

$$y_{ij} = \begin{cases} 1 & j \in J; \\ 0 & \text{otherwise.} \end{cases} \tag{A.27}$$

For instance, if a multi-class classification problem contains three classes $A$, $B$, and $C$, then the labels $\mathbf{y}_i$ can be encoded in $\{A : (1, 0, 0); B : (0, 1, 0); C : (0, 0, 1)\}$.

Assume that our neural network is parametrized by $\theta$. For true label vectors $\mathbf{y}_i$ and predictions

$$\hat{\mathbf{y}}_i = p_\theta(\mathbf{y}_i \mid \mathbf{x}_i) = \sum_{j=1}^{k} y_{ij} p_\theta(y_{ij} \mid \mathbf{x}_i). \tag{A.28}$$

Hence, the *cross-entropy loss* would be

$$\mathrm{CE}(\mathbf{y}, \hat{\mathbf{y}}) = -\sum_{i=1}^{n} \mathbf{y}_i \log \hat{\mathbf{y}}_i = -\sum_{i=1}^{n} \sum_{j=1}^{k} y_{ij} \log p_\theta(y_{ij} \mid \mathbf{x}_i). \tag{A.29}$$

On the other side, we can also approach the problem through maximum likelihood estimation. To begin with, let's quickly introduce a $k$-class multinoulli distribution. It is an extension of the Bernoulli distribution from binary class to multi-class. If a random variable $\mathbf{z} = (z_1, \ldots, z_k)$ follows a $k$-class *multinoulli distribution* with probabilities $\mathbf{p} = (p_1, \ldots, p_k)$, i.e.,

$$p(\mathbf{z}) = p(z_1, \ldots, z_k) = \mathrm{Multi}(p_1, \ldots, p_k), \text{ where } \sum_{i=1}^{k} p_i = 1, \tag{A.30}$$

then the joint probability mass function(p.m.f.) of $\mathbf{z}$ is

$$\mathbf{p}^{\mathbf{z}} = \prod_{j=1}^{k} p_j^{z_j}. \tag{A.31}$$

It can be seen that the label of each data example, $\mathbf{y}_i$, is following a $k$-class multinoulli distribution with probabilities $\boldsymbol{\pi} = (\pi_1, \ldots, \pi_k)$. Therefore, the joint p.m.f. of each data example $\mathbf{y}_i$ is ß$^{\mathbf{y}_i} = \prod_{j=1}^{k} \pi_j^{y_{ij}}$. Hence, the log-likelihood function would be

$$l(\theta) = \log L(\theta) = \log \prod_{i=1}^{n} \boldsymbol{\pi}^{\mathbf{y}_i} = \log \prod_{i=1}^{n} \prod_{j=1}^{k} \pi_j^{y_{ij}} = \sum_{i=1}^{n} \sum_{j=1}^{k} y_{ij} \log \pi_j. \tag{A.32}$$

Since in maximum likelihood estimation, we maximizing the objective function $l(\theta)$ by having $\pi_j = p_\theta(y_{ij} \mid \mathbf{x}_i)$. Therefore, for any multi-class classification, maximizing the above log-likelihood function $l(\theta)$ is equivalent to minimizing the CE loss $\mathrm{CE}(y, \hat{y})$.

To test the above proof, let's apply the built-in measure `NegativeLogLikelihood`. Using the same `labels` and `preds` as in the earlier example, we will get the same numerical loss as the previous example up to the 5 decimal place.

```
# Implementation of cross-entropy loss in PyTorch combines `nn.LogSoftmax()`
# and `nn.NLLLoss()`
nll_loss = NLLLoss()
loss = nll_loss(torch.log(preds), labels)
loss
```

```
tensor(0.9486)
```

## A.11.6 Summary

- Information theory is a field of study about encoding, decoding, transmitting, and manipulating information.
- Entropy is the unit to measure how much information is presented in different signals.
- KL divergence can also measure the divergence between two distributions.
- Cross-entropy can be viewed as an objective function of multi-class classification. Minimizing cross-entropy loss is equivalent to maximizing the log-likelihood function.

## A.11.7 Exercises

1. Verify that the card examples from the first section indeed have the claimed entropy.
2. Show that the KL divergence $D(p\|q)$ is nonnegative for all distributions $p$ and $q$. Hint: use Jensen's inequality, i.e., use the fact that $-\log x$ is a convex function.
3. Let's compute the entropy from a few data sources:
   - Assume that you are watching the output generated by a monkey at a typewriter. The monkey presses any of the 44 keys of the typewriter at random (you can assume that it has not discovered any special keys or the shift key yet). How many bits of randomness per character do you observe?

- Being unhappy with the monkey, you replaced it by a drunk typesetter. It is able to generate words, albeit not coherently. Instead, it picks a random word out of a vocabulary of $2,000$ words. Let's assume that the average length of a word is $4.5$ letters in English. How many bits of randomness per character do you observe now?
- Still being unhappy with the result, you replace the typesetter by a high quality language model. The language model can currently obtain a perplexity as low as $15$ points per word. The character *perplexity* of a language model is defined as the inverse of the geometric mean of a set of probabilities, each probability is corresponding to a character in the word. To be specific, if the length of a given word is $l$, then $\mathrm{PPL}(\text{word}) = [\prod_i p(\text{character}_i)]^{-\frac{1}{l}} = \exp\left[-\frac{1}{l}\sum_i \log p(\text{character}_i)\right]$. Assume that the test word has 4.5 letters, how many bits of randomness per character do you observe now?

4. Explain intuitively why $I(X,Y) = H(X) - H(X \mid Y)$. Then, show this is true by expressing both sides as an expectation with respect to the joint distribution.

5. What is the KL Divergence between the two Gaussian distributions $\mathcal{N}(\mu_1, \sigma_1^2)$ and $\mathcal{N}(\mu_2, \sigma_2^2)$?

Discussions[288].

# B Tools for Deep Learning

To get the most out of *Dive into Deep Learning*, we will talk you through different tools in this appendix, such as for running and contributing to this interactive open-source book.

## B.1 Using Jupyter Notebooks

This section describes how to edit and run the code in each section of this book using the Jupyter Notebook. Make sure you have installed Jupyter and downloaded the code as described in *Installation* (page xxxiv). If you want to know more about Jupyter see the excellent tutorial in their documentation[289].

### B.1.1 Editing and Running the Code Locally

Suppose that the local path of the book's code is `xx/yy/d2l-en/`. Use the shell to change the directory to this path (`cd xx/yy/d2l-en`) and run the command `jupyter notebook`. If your browser does not do this automatically, open http://localhost:8888 and you will see the interface of Jupyter and all the folders containing the code of the book, as shown in Fig. B.1.

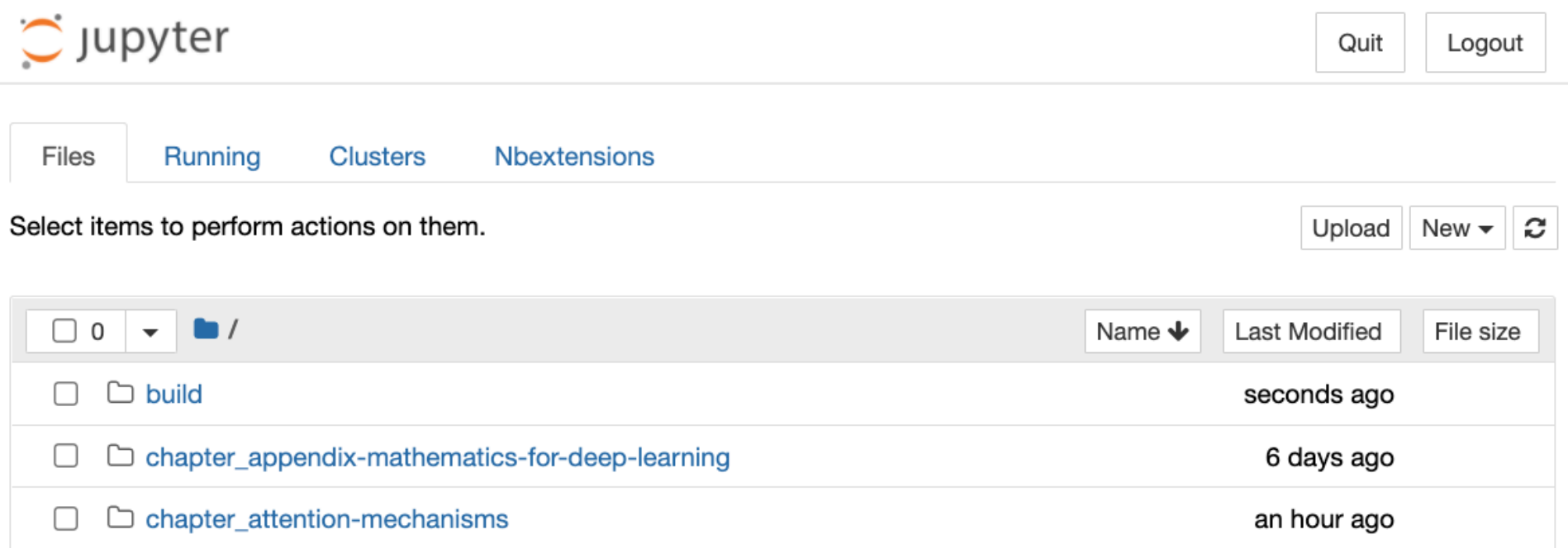

Fig. B.1 The folders containing the code of this book.

You can access the notebook files by clicking on the folder displayed on the webpage. They usually have the suffix ".ipynb". For the sake of brevity, we create a temporary "test.ipynb" file. The content displayed after you click it is shown in Fig. B.2. This notebook includes a markdown cell and a code cell. The content in the markdown cell includes "This Is a Title" and "This is text.". The code cell contains two lines of Python code.

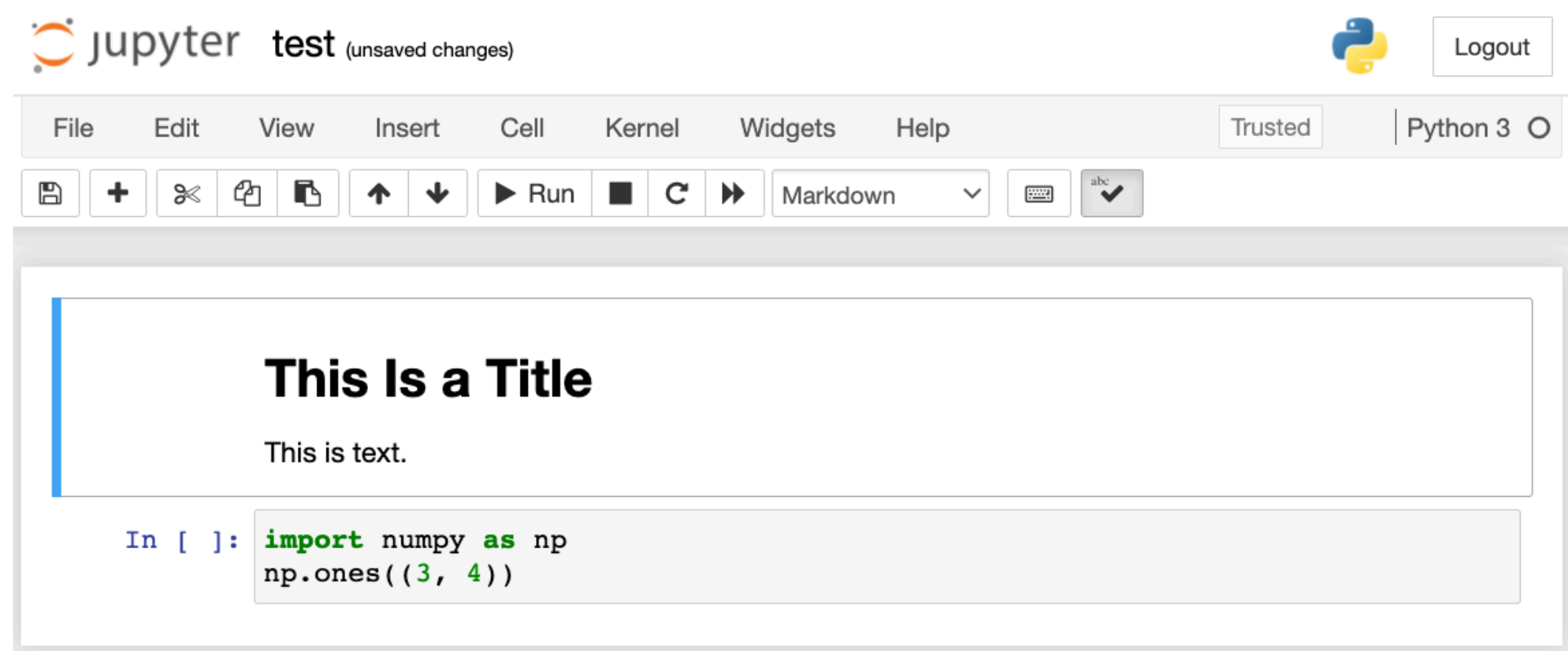

Fig. B.2 Markdown and code cells in the "text.ipynb" file.

Double click on the markdown cell to enter edit mode. Add a new text string "Hello world." at the end of the cell, as shown in Fig. B.3.

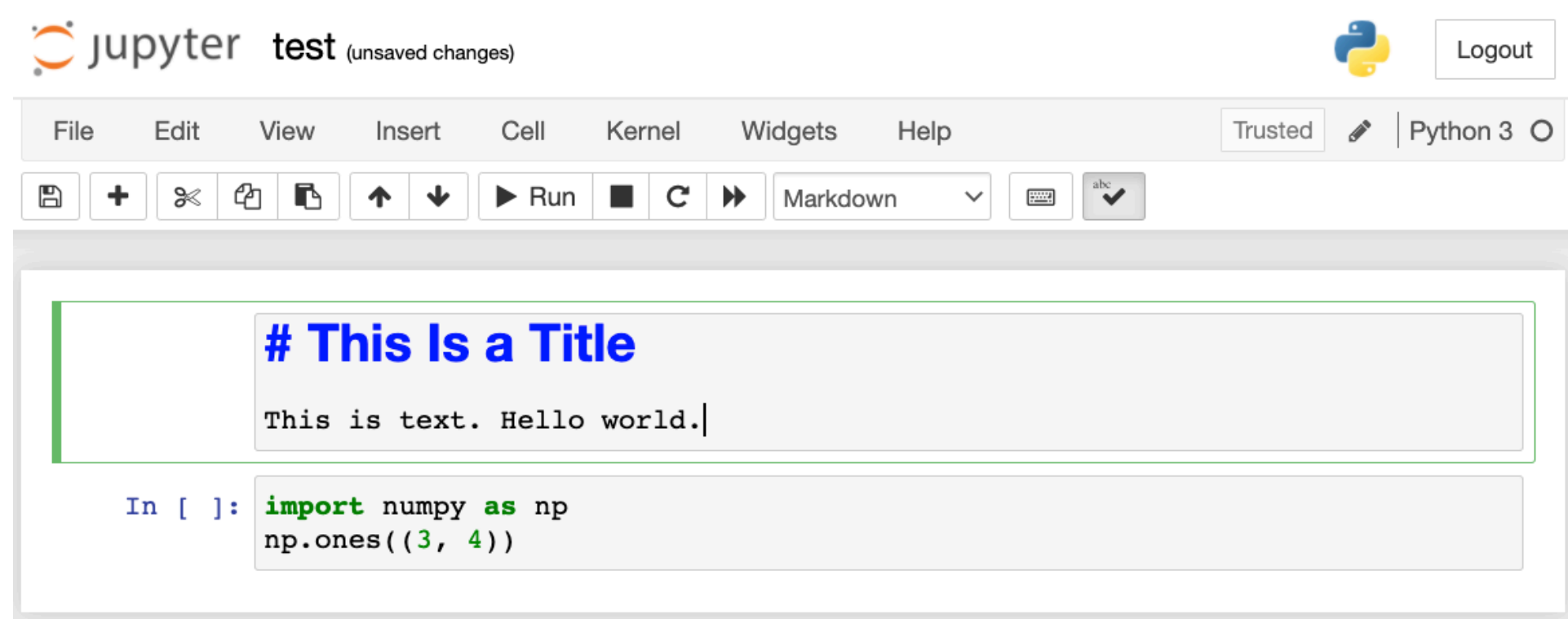

Fig. B.3 Edit the markdown cell.

As demonstrated in Fig. B.4, click "Cell" → "Run Cells" in the menu bar to run the edited cell.

After running, the markdown cell is shown in Fig. B.5.

Next, click on the code cell. Multiply the elements by 2 after the last line of code, as shown in Fig. B.6.

You can also run the cell with a shortcut ("Ctrl + Enter" by default) and obtain the output result from Fig. B.7.

When a notebook contains more cells, we can click "Kernel" → "Restart & Run All" in the menu bar to run all the cells in the entire notebook. By clicking "Help" → "Edit Keyboard Shortcuts" in the menu bar, you can edit the shortcuts according to your preferences.

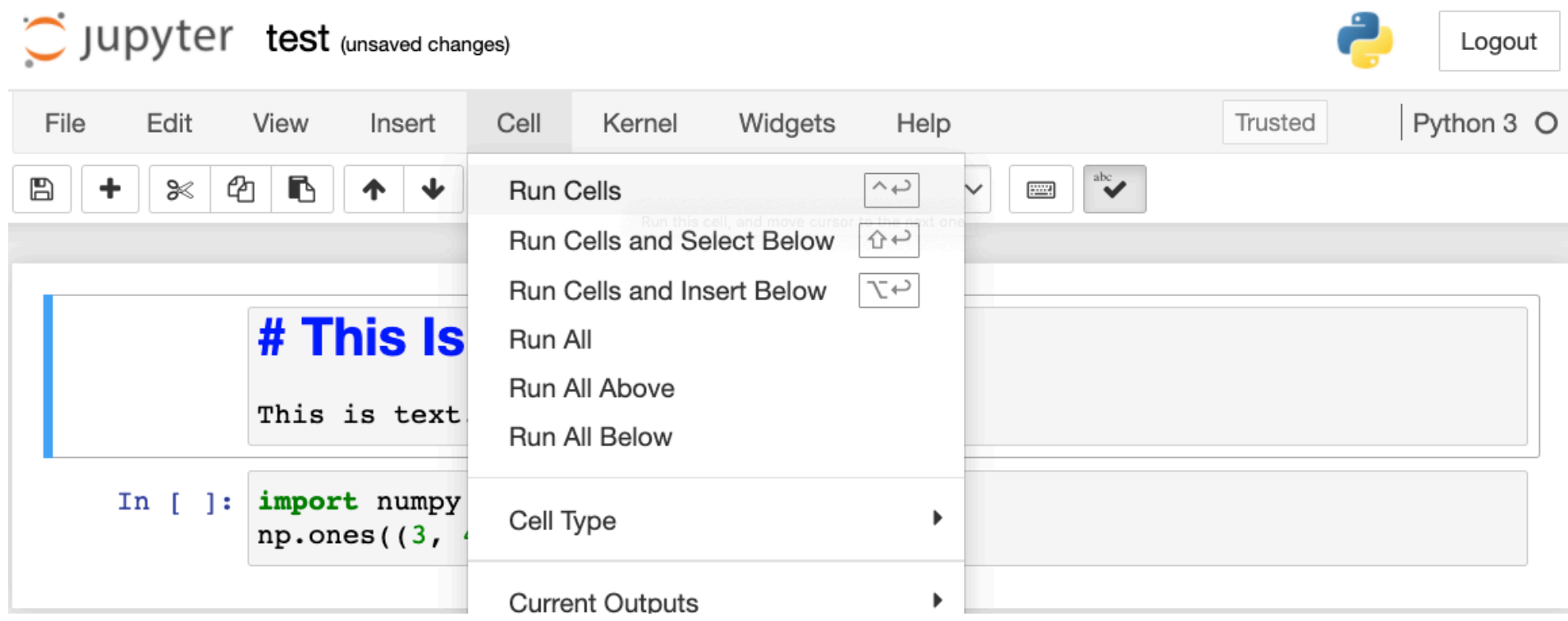

Fig. B.4 Run the cell.

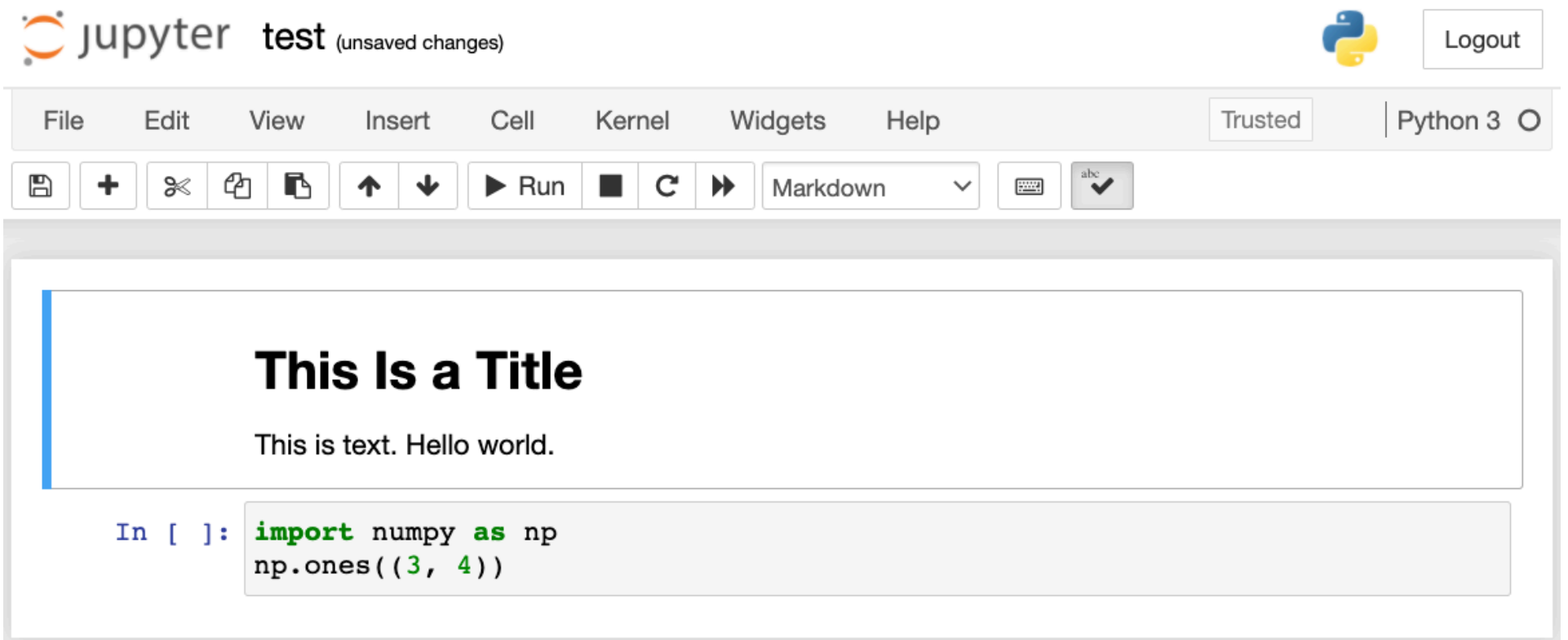

Fig. B.5 The markdown cell after running.

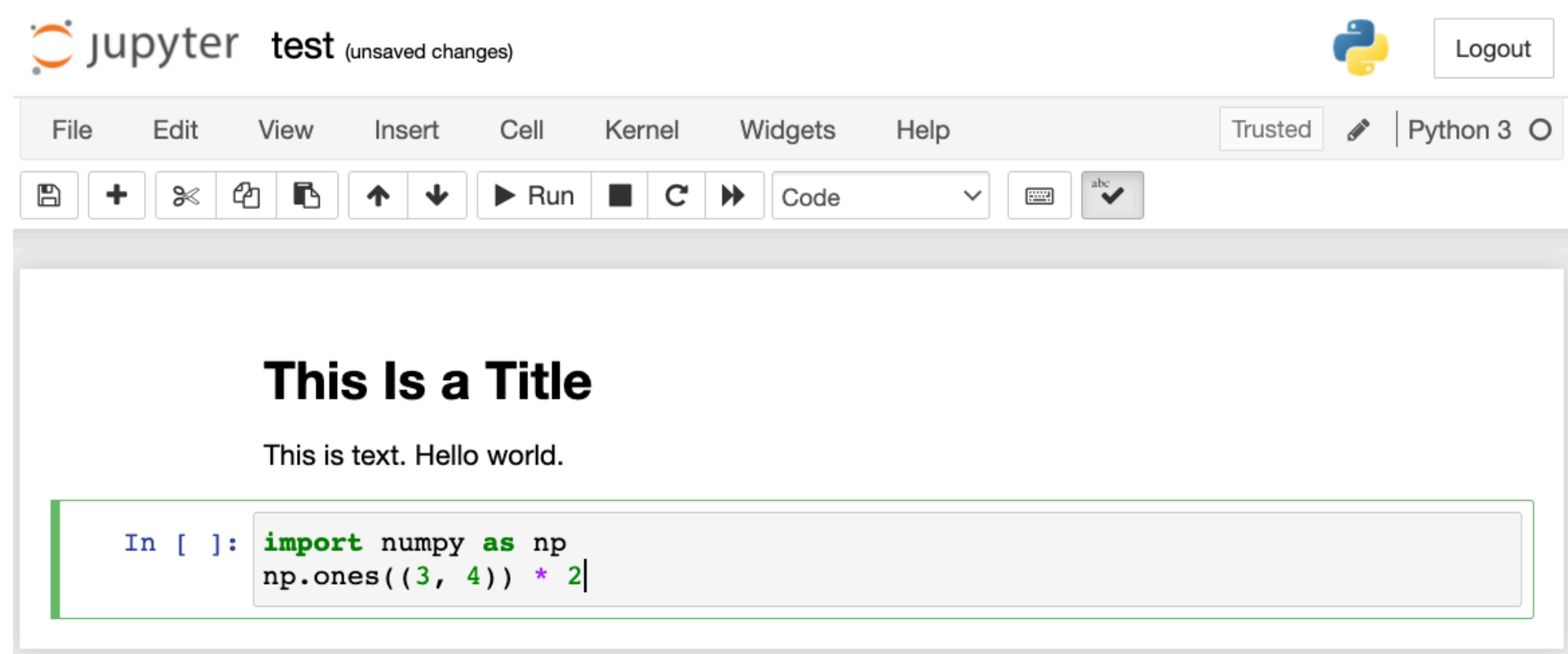

Fig. B.6 Edit the code cell.

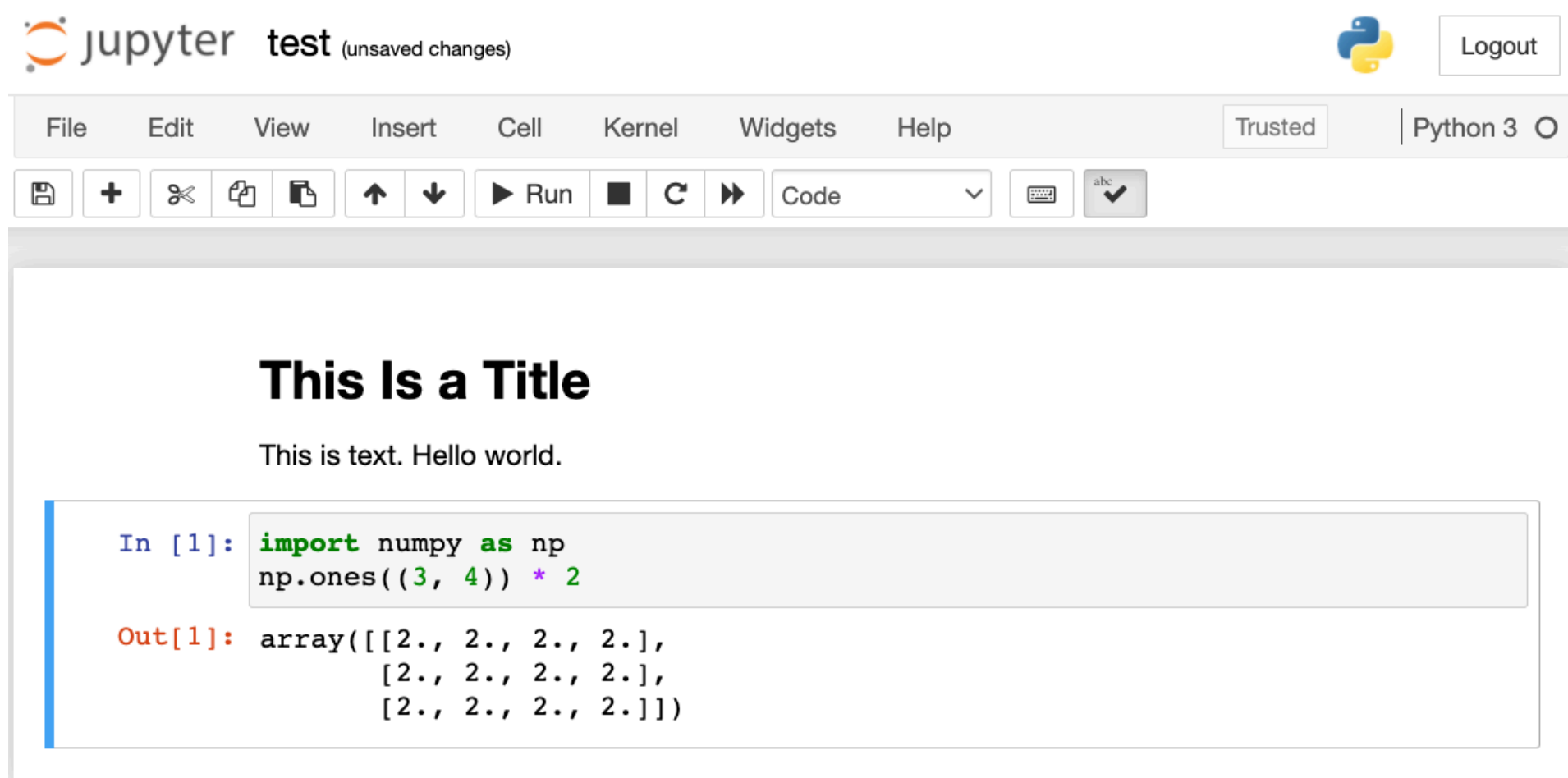

Fig. B.7 Run the code cell to obtain the output.

## B.1.2 Advanced Options

Beyond local editing two things are quite important: editing the notebooks in the markdown format and running Jupyter remotely. The latter matters when we want to run the code on a faster server. The former matters since Jupyter's native ipynb format stores a lot of auxiliary data that is irrelevant to the content, mostly related to how and where the code is run. This is confusing for Git, making reviewing contributions very difficult. Fortunately there is an alternative—native editing in the markdown format.

### Markdown Files in Jupyter

If you wish to contribute to the content of this book, you need to modify the source file (md file, not ipynb file) on GitHub. Using the notedown plugin we can modify notebooks in the md format directly in Jupyter.

First, install the notedown plugin, run the Jupyter Notebook, and load the plugin:

```
pip install d2l-notedown  # You may need to uninstall the original notedown.
jupyter notebook --NotebookApp.contents_manager_class='notedown.
↪NotedownContentsManager'
```

You may also turn on the notedown plugin by default whenever you run the Jupyter Notebook. First, generate a Jupyter Notebook configuration file (if it has already been generated, you can skip this step).

```
jupyter notebook --generate-config
```

Then, add the following line to the end of the Jupyter Notebook configuration file (for Linux or macOS, usually in the path `~/.jupyter/jupyter_notebook_config.py`):

```
c.NotebookApp.contents_manager_class = 'notedown.NotedownContentsManager'
```

After that, you only need to run the `jupyter notebook` command to turn on the notedown plugin by default.

### Running Jupyter Notebooks on a Remote Server

Sometimes, you may want to run Jupyter notebooks on a remote server and access it through a browser on your local computer. If Linux or macOS is installed on your local machine (Windows can also support this function through third-party software such as PuTTY), you can use port forwarding:

```
ssh myserver -L 8888:localhost:8888
```

The above string `myserver` is the address of the remote server. Then we can use http://localhost:8888 to access the remote server `myserver` that runs Jupyter notebooks. We will detail on how to run Jupyter notebooks on AWS instances later in this appendix.

### Timing

We can use the `ExecuteTime` plugin to time the execution of each code cell in Jupyter notebooks. Use the following commands to install the plugin:

```
pip install jupyter_contrib_nbextensions
jupyter contrib nbextension install --user
jupyter nbextension enable execute_time/ExecuteTime
```

## B.1.3 Summary

- Using the Jupyter Notebook tool, we can edit, run, and contribute to each section of the book.
- We can run Jupyter notebooks on remote servers using port forwarding.

### B.1.4 Exercises

1. Edit and run the code in this book with the Jupyter Notebook on your local machine.
2. Edit and run the code in this book with the Jupyter Notebook *remotely* via port forwarding.
3. Compare the running time of the operations $\mathbf{A}^\top\mathbf{B}$ and $\mathbf{A}\mathbf{B}$ for two square matrices in $\mathbb{R}^{1024\times1024}$. Which one is faster?

Discussions[290].

## B.2 Using Amazon SageMaker

Deep learning applications may demand so much computational resource that easily goes beyond what your local machine can offer. Cloud computing services allow you to run GPU-intensive code of this book more easily using more powerful computers. This section will introduce how to use Amazon SageMaker to run the code of this book.

### B.2.1 Signing Up

First, we need to sign up an account at https://aws.amazon.com/. For additional security, using two-factor authentication is encouraged. It is also a good idea to set up detailed billing and spending alerts to avoid any surprise, e.g., when forgetting to stop running instances. After logging into your AWS account, go to your console[291] and search for "Amazon SageMaker" (see Fig. B.1), then click it to open the SageMaker panel.

Fig. B.1 Search for and open the SageMaker panel.

### B.2.2 Creating a SageMaker Instance

Next, let's create a notebook instance as described in Fig. B.2.

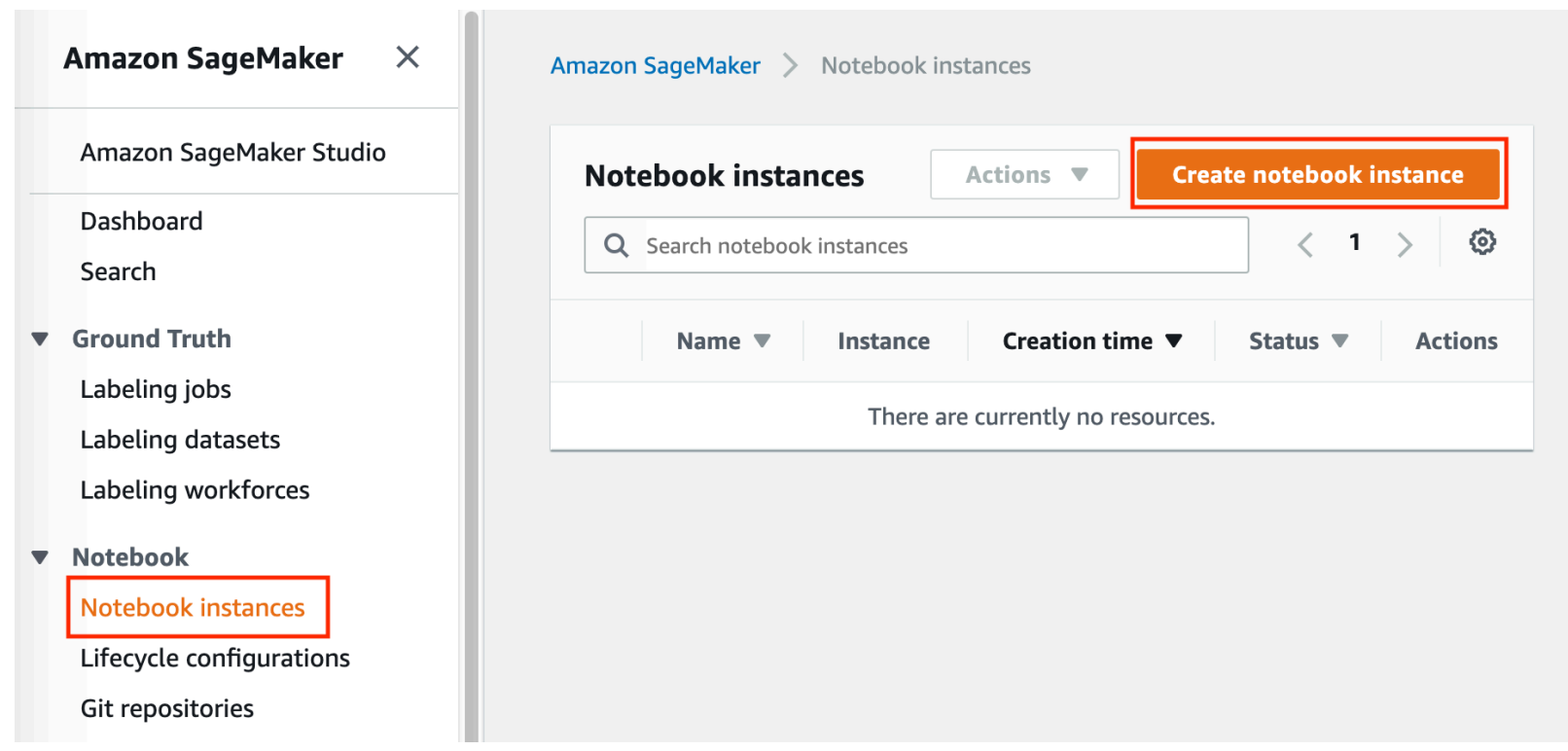

Fig. B.2 Create a SageMaker instance.

SageMaker provides multiple instance types[292] with varying computational power and prices. When creating a notebook instance, we can specify its name and type. In Fig. B.3, we choose `ml.p3.2xlarge`: with one Tesla V100 GPU and an 8-core CPU, this instance is powerful enough for most of the book.

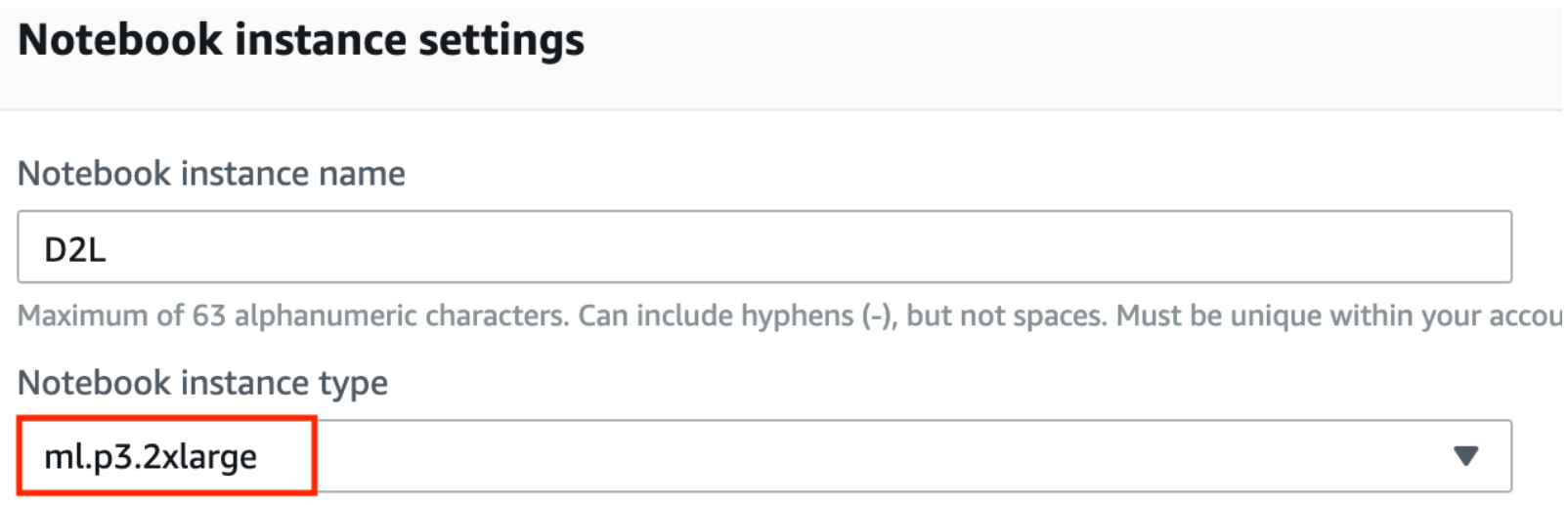

Fig. B.3 Choose the instance type.

The entire book in the ipynb format for running with SageMaker is available at https://github.com/d2l-ai/d2l-pytorch-sagemaker. We can specify this GitHub repository URL (Fig. B.4) to allow SageMaker to clone it when creating the instance.

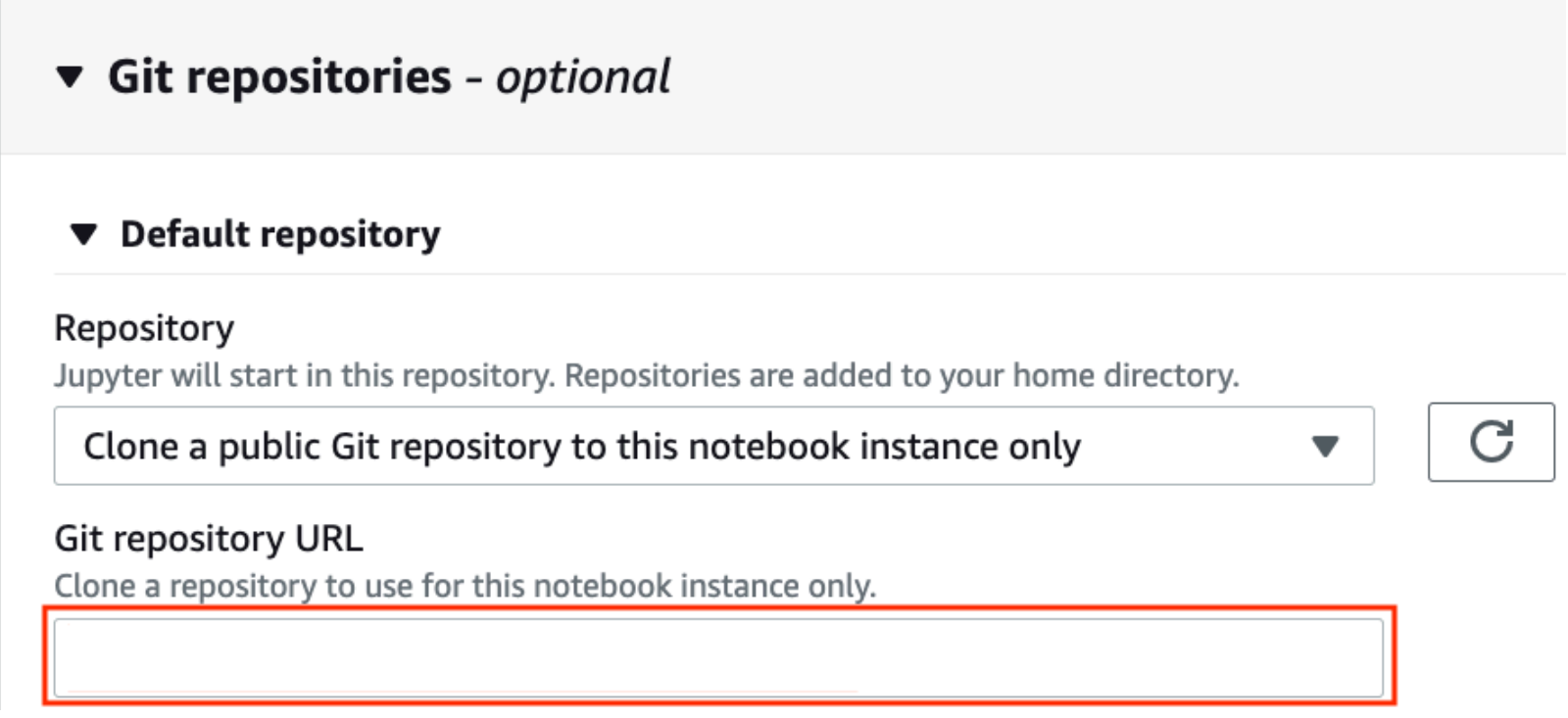

Fig. B.4 Specify the GitHub repository.

### B.2.3 Running and Stopping an Instance

Creating an instance may take a few minutes. When it is ready, click on the "Open Jupyter" link next to it (Fig. B.5) so you can edit and run all the Jupyter notebooks of this book on this instance (similar to steps in Section B.1).

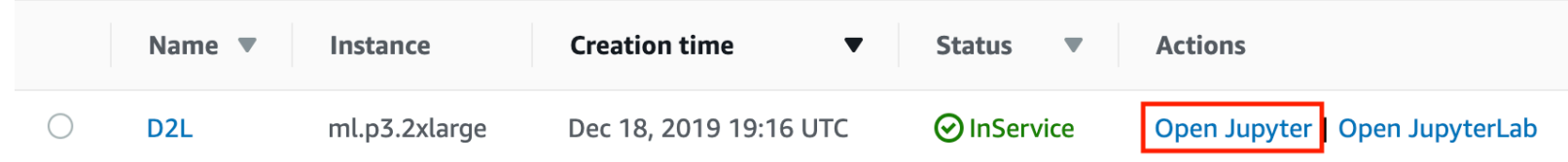

Fig. B.5 Open Jupyter on the created SageMaker instance.

After finishing your work, do not forget to stop the instance to avoid being charged further (Fig. B.6).

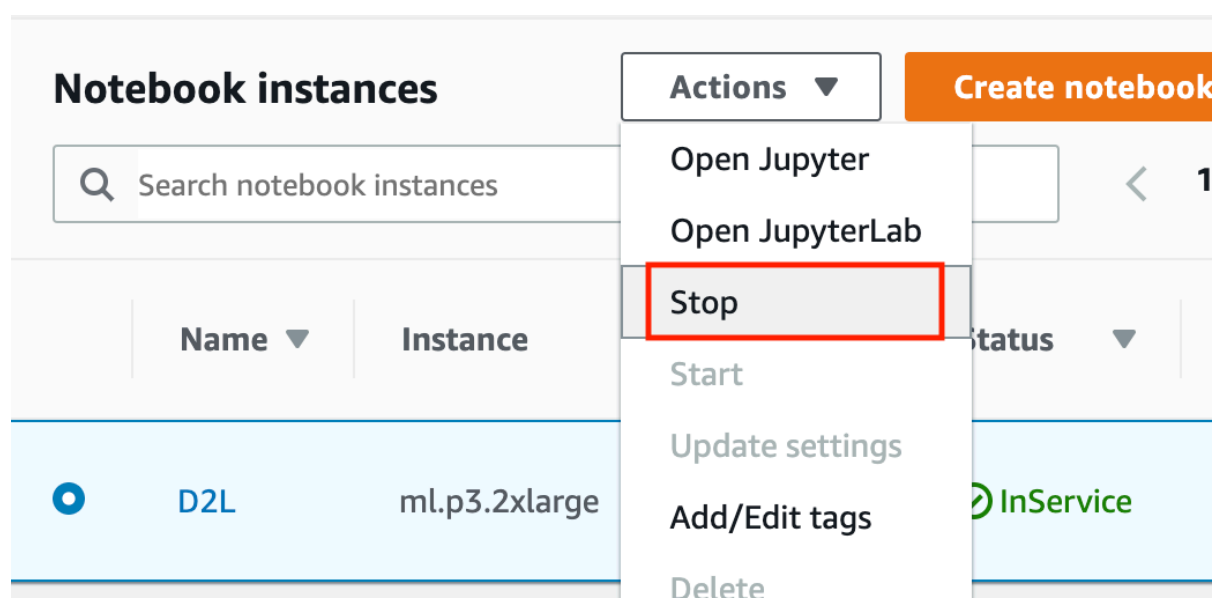

Fig. B.6 Stop a SageMaker instance.

### B.2.4 Updating Notebooks

Notebooks of this open-source book will be regularly updated in the d2l-ai/d2l-pytorch-sagemaker[293] repository on GitHub. To update to the latest version, you may open a terminal on the SageMaker instance (Fig. B.7).

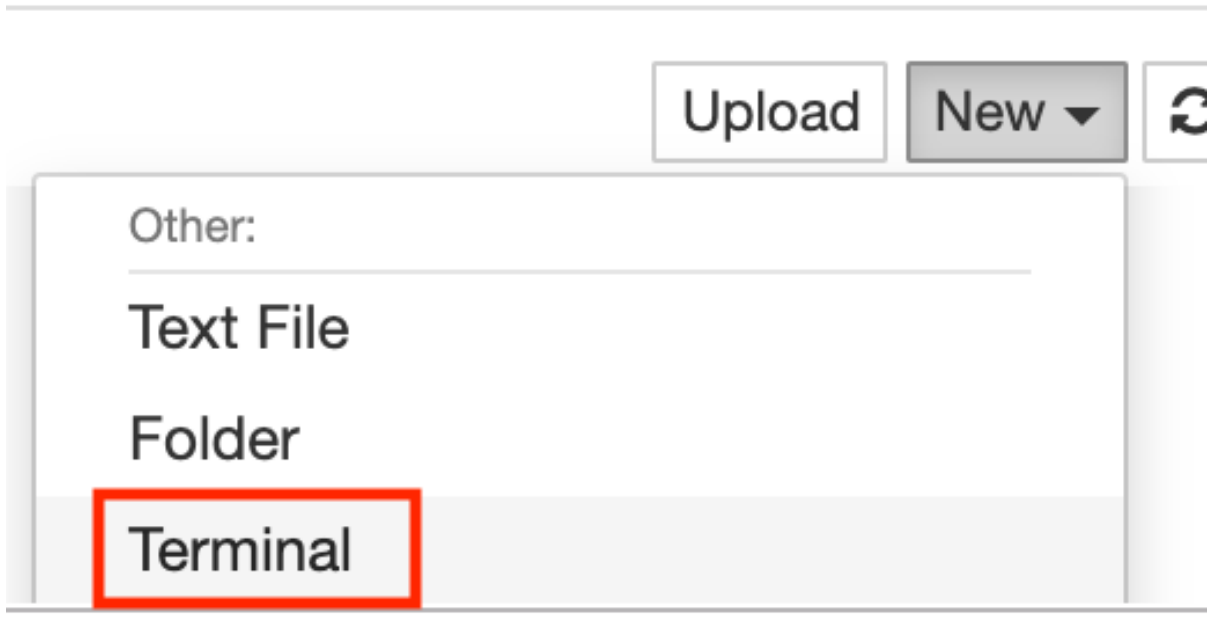

Fig. B.7 Open a terminal on the SageMaker instance.

You may wish to commit your local changes before pulling updates from the remote repos-itory. Otherwise, simply discard all your local changes with the following commands in the terminal:

```
cd SageMaker/d2l-pytorch-sagemaker/
git reset --hard
git pull
```

### B.2.5 Summary

- We can create a notebook instance using Amazon SageMaker to run GPU-intensive code of this book.
- We can update notebooks via the terminal on the Amazon SageMaker instance.

### B.2.6 Exercises

1. Edit and run any section that requires a GPU using Amazon SageMaker.
2. Open a terminal to access the local directory that hosts all the notebooks of this book.

Discussions[294].

294

## B.3 Using AWS EC2 Instances

In this section, we will show you how to install all libraries on a raw Linux machine. Recall that in Section B.2 we discussed how to use Amazon SageMaker, while building an instance by yourself costs less on AWS. The walkthrough includes three steps:

1. Request for a GPU Linux instance from AWS EC2.
2. Install CUDA (or use an Amazon Machine Image with preinstalled CUDA).
3. Install the deep learning framework and other libraries for running the code of the book.

This process applies to other instances (and other clouds), too, albeit with some minor modifications. Before going forward, you need to create an AWS account, see Section B.2 for more details.

### B.3.1 Creating and Running an EC2 Instance

After logging into your AWS account, click "EC2" (Fig. B.1) to go to the EC2 panel.

Fig. B.2 shows the EC2 panel.

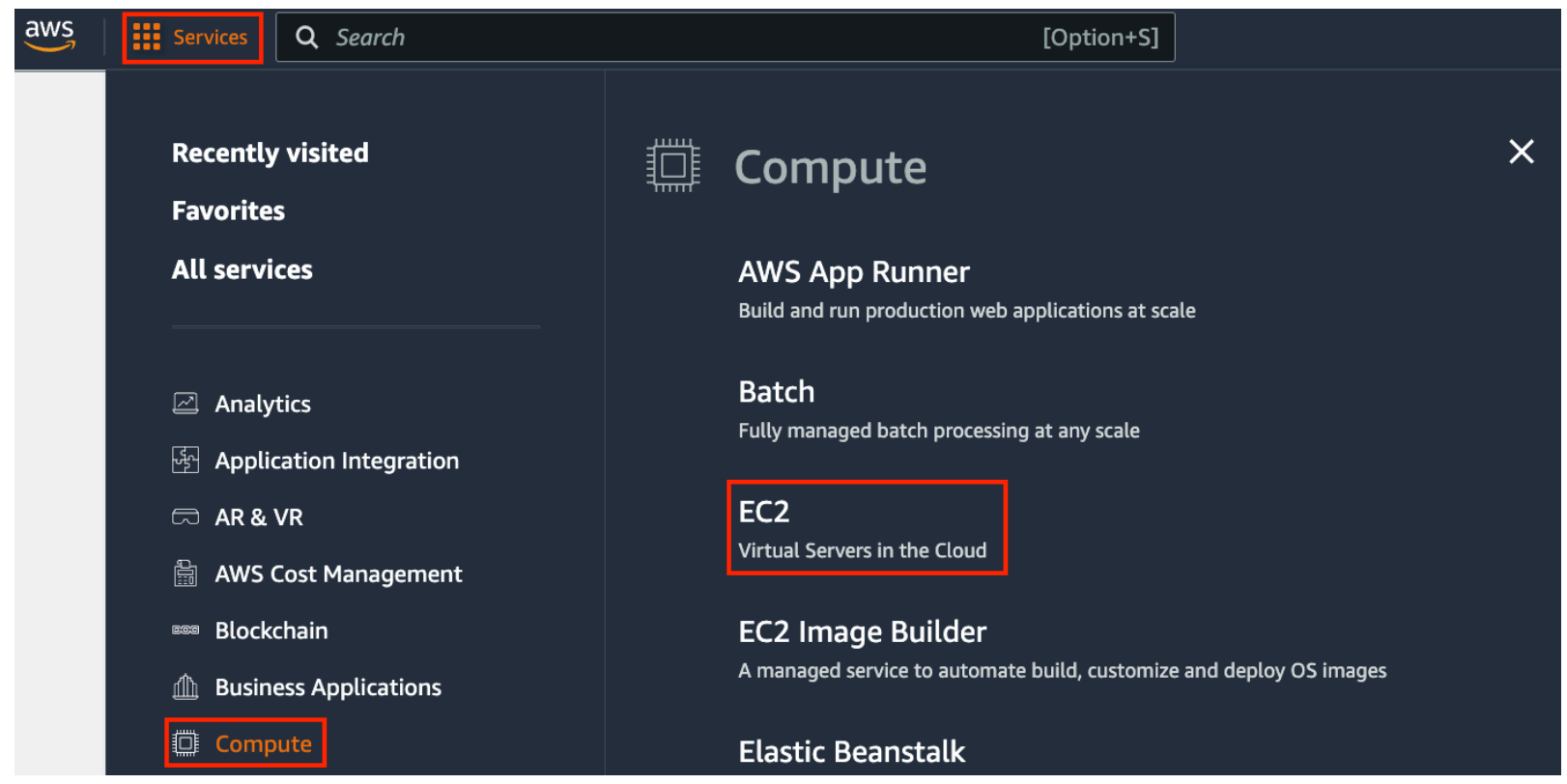

Fig. B.1 Open the EC2 console.

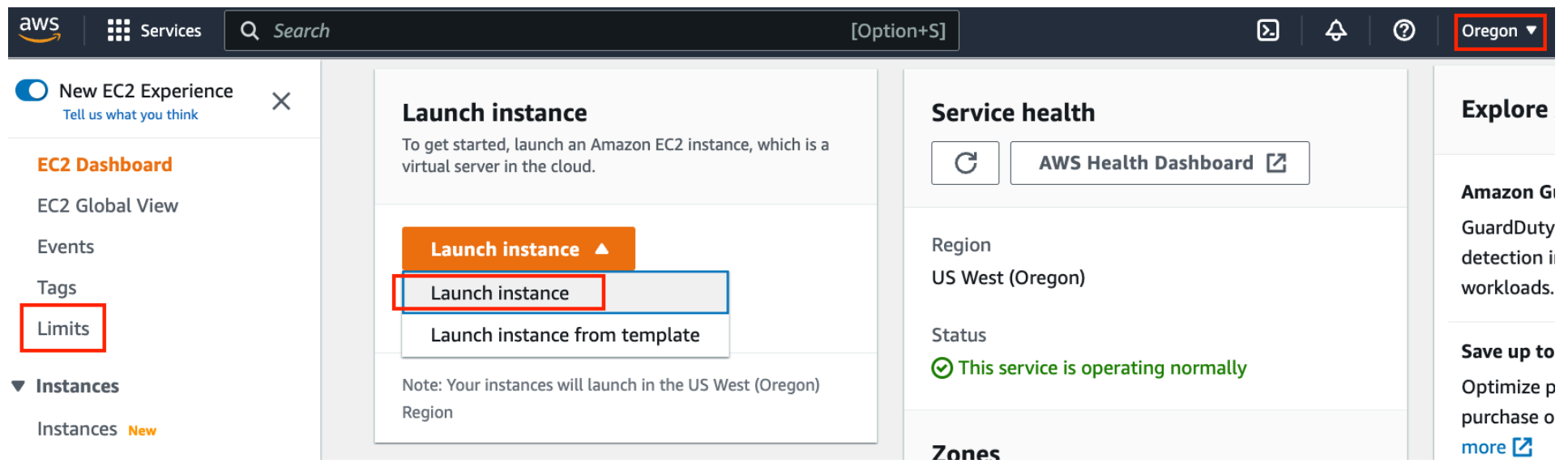

Fig. B.2 The EC2 panel.

## Presetting Location

Select a nearby data center to reduce latency, e.g., “Oregon” (marked by the red box in the top-right of Fig. B.2). If you are located in China, you can select a nearby Asia Pacific region, such as Seoul or Tokyo. Please note that some data centers may not have GPU instances.

## Increasing Limits

Before choosing an instance, check if there are quantity restrictions by clicking the “Limits” label in the bar on the left as shown in Fig. B.2. Fig. B.3 shows an example of such a limitation. The account currently cannot open “p2.xlarge” instances according to the region. If you need to open one or more instances, click on the “Request limit increase” link to apply for a higher instance quota. Generally, it takes one business day to process an application.

## Launching an Instance

Next, click the “Launch Instance” button marked by the red box in Fig. B.2 to launch your instance.

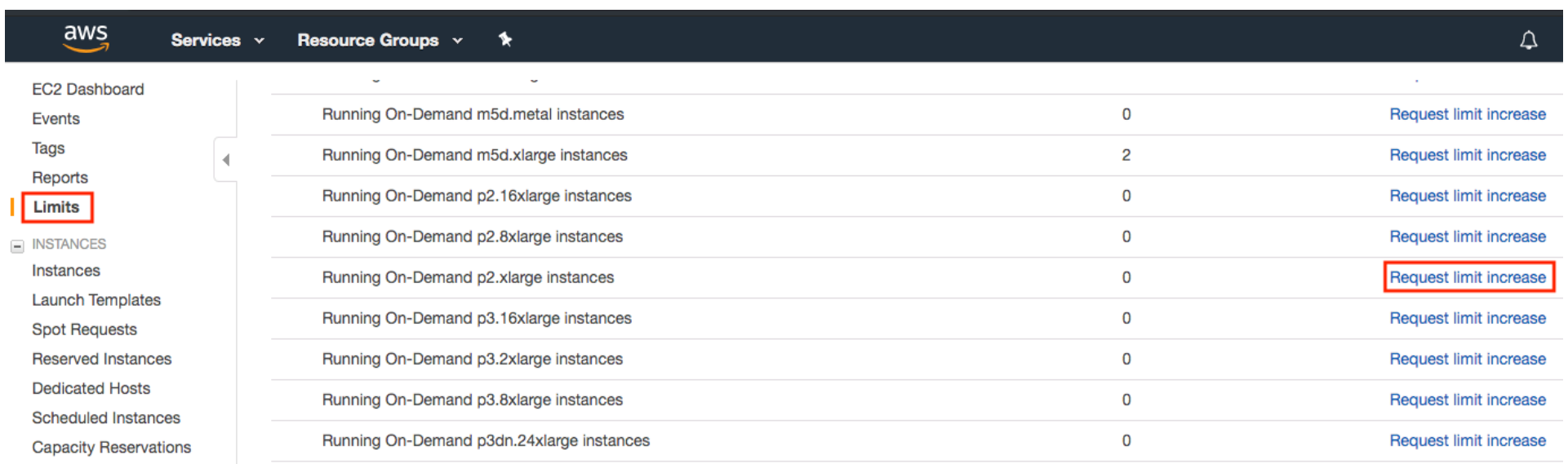

Fig. B.3 Instance quantity restrictions.

We begin by selecting a suitable Amazon Machine Image (AMI). Select an Ubuntu instance (Fig. B.4).

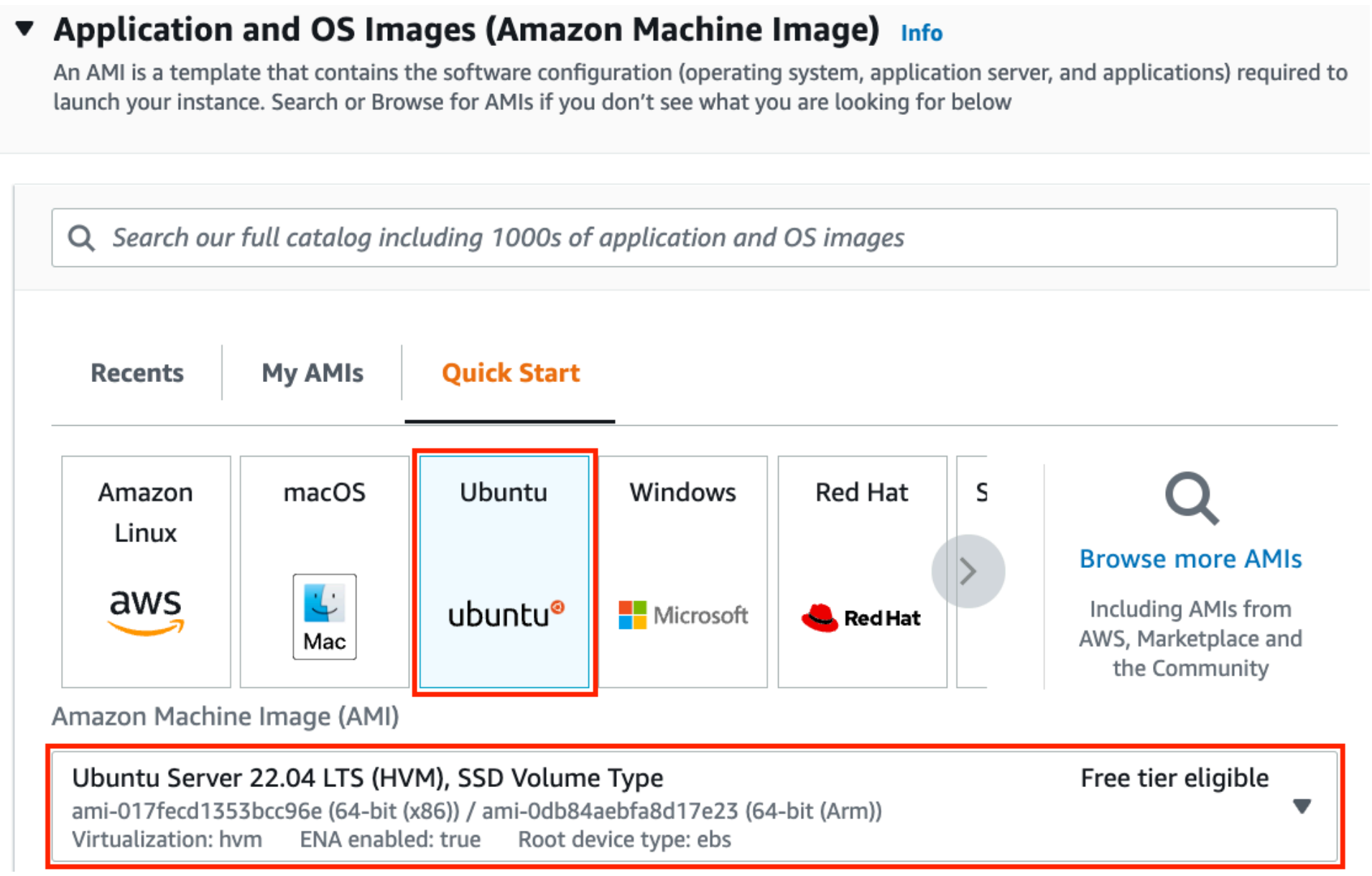

Fig. B.4 Choose an AMI.

EC2 provides many different instance configurations to choose from. This can sometimes feel overwhelming to a beginner. Table B.1 lists different suitable machines.

Table B.1: Different EC2 instance types

| Name | GPU | Notes |
|---|---|---|
| g2 | Grid K520 | ancient |
| p2 | Kepler K80 | old but often cheap as spot |
| g3 | Maxwell M60 | good trade-off |
| p3 | Volta V100 | high performance for FP16 |
| p4 | Ampere A100 | high performance for large-scale training |
| g4 | Turing T4 | inference optimized FP16/INT8 |

All these servers come in multiple flavors indicating the number of GPUs used. For example, a p2.xlarge has 1 GPU and a p2.16xlarge has 16 GPUs and more memory. For more details, see the AWS EC2 documentation[295] or a summary page[296]. For the purpose of illustration, a p2.xlarge will suffice (marked in the red box of Fig. B.5).

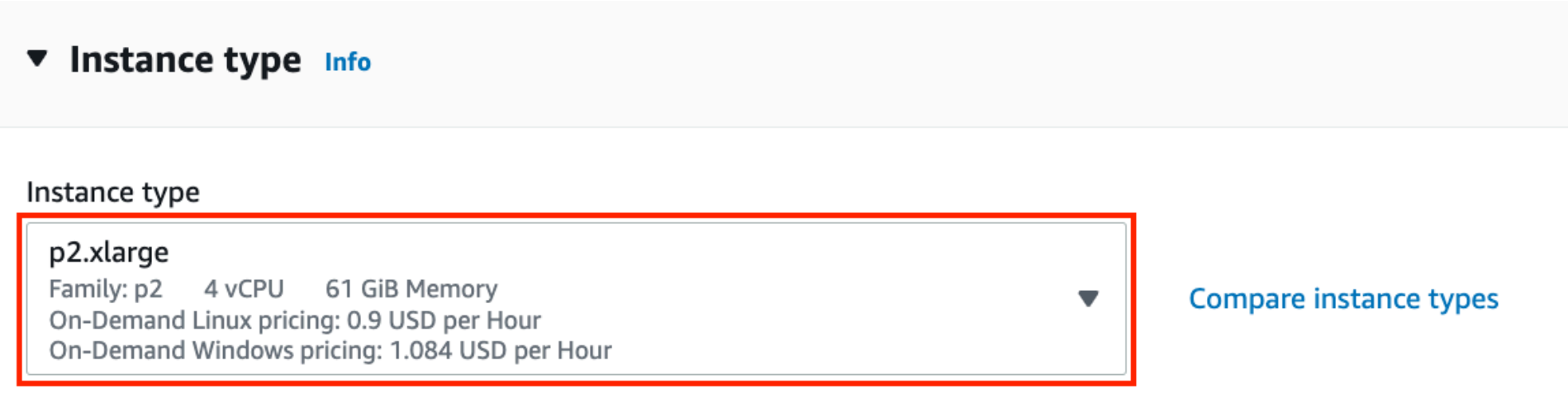

Fig. B.5 Choose an instance.

Note that you should use a GPU-enabled instance with suitable drivers and a GPU-enabled deep learning framework. Otherwise you will not see any benefit from using GPUs.

We go on to select the key pair used to access the instance. If you do not have a key pair, click "Create new key pair" in Fig. B.6 to generate a key pair. Subsequently, you can select the previously generated key pair. Make sure that you download the key pair and store it in a safe location if you generated a new one. This is your only way to SSH into the server.

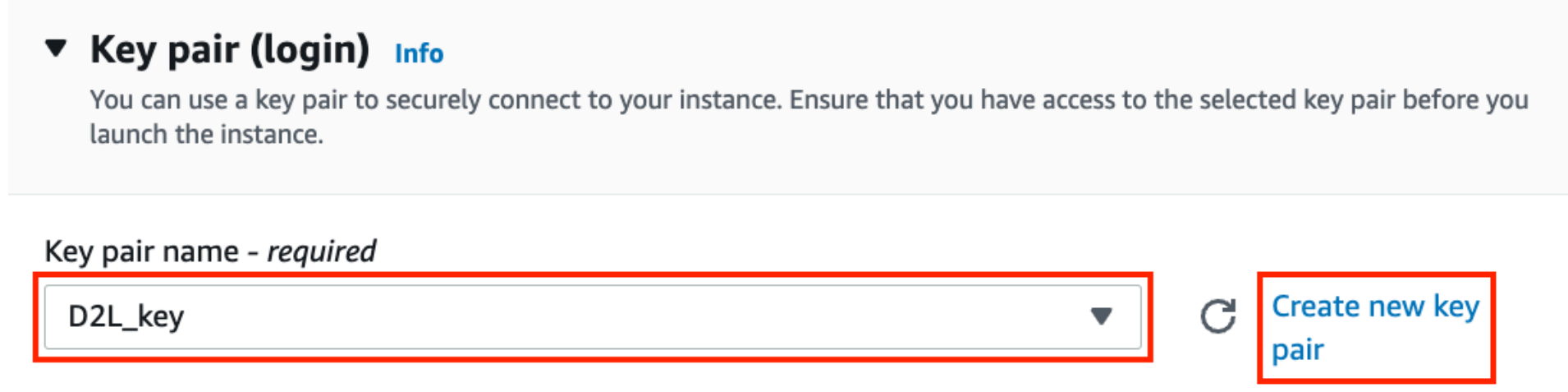

Fig. B.6 Select a key pair.

In this example, we will keep the default configurations for "Network settings" (click the "Edit" button to configure items such as the subnet and security groups). We just increase the default hard disk size to 64 GB (Fig. B.7). Note that CUDA by itself already takes up 4 GB.

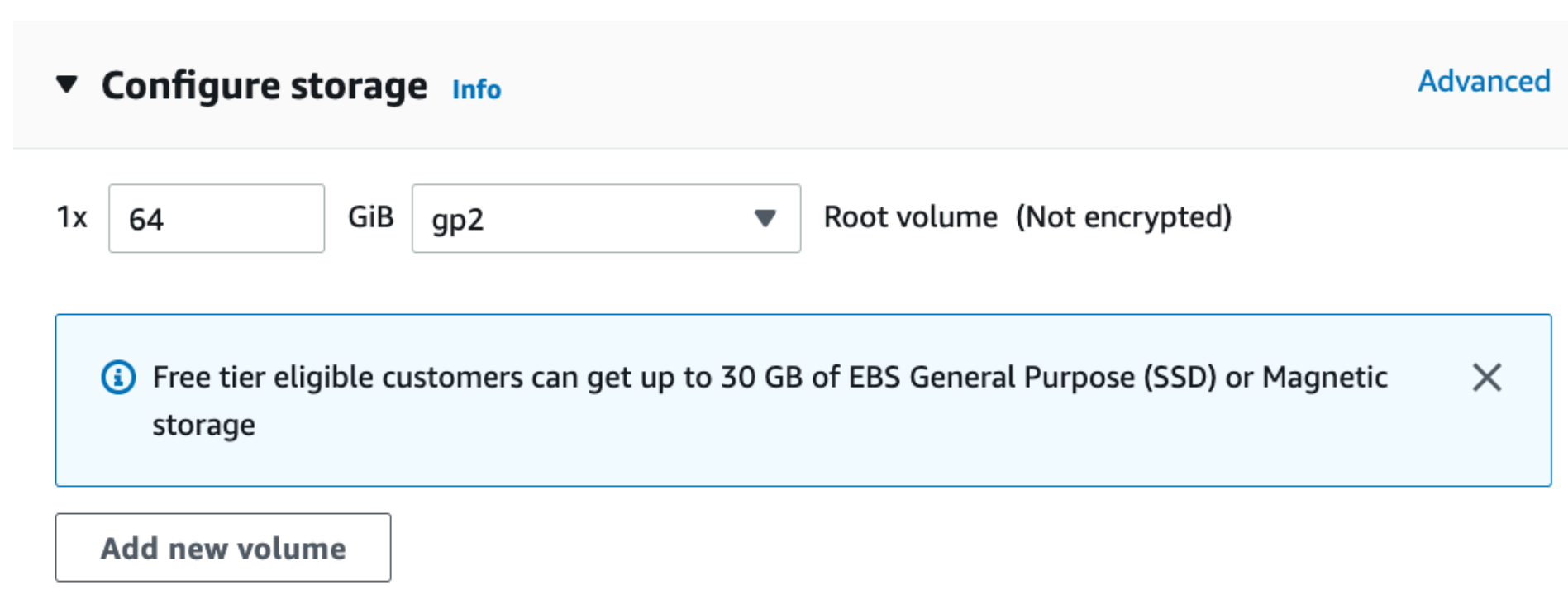

Fig. B.7 Modify the hard disk size.

Click "Launch Instance" to launch the created instance. Click the instance ID shown in Fig. B.8 to view the status of this instance.

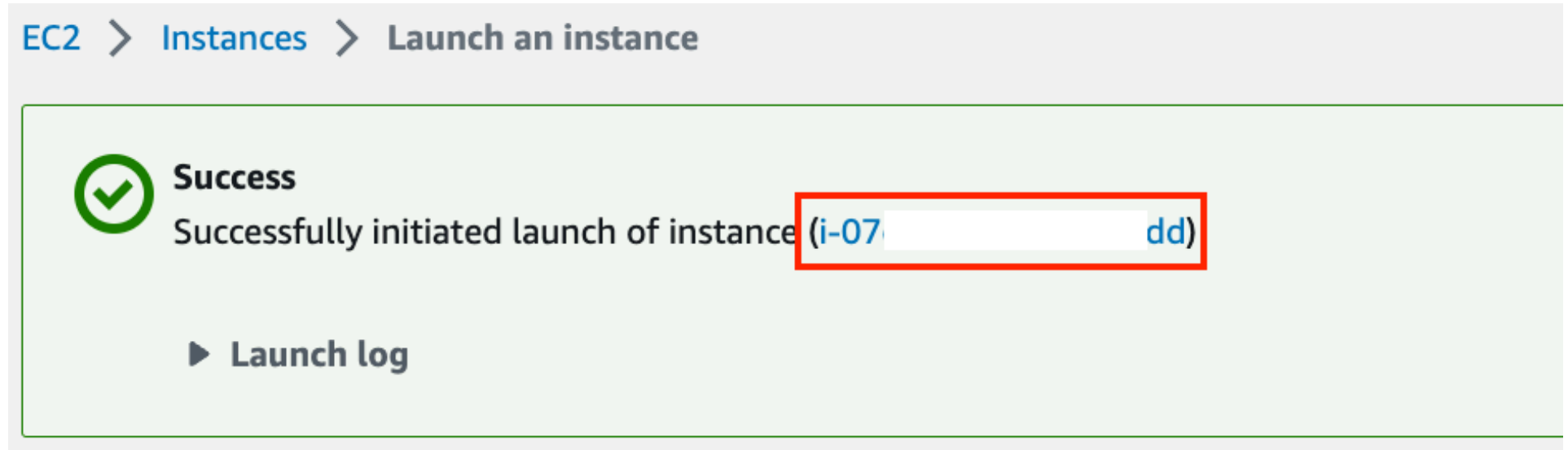

Fig. B.8 Click the instance ID.

## Connecting to the Instance

As shown in Fig. B.9, after the instance state turns green, right-click the instance and select `Connect` to view the instance access method.

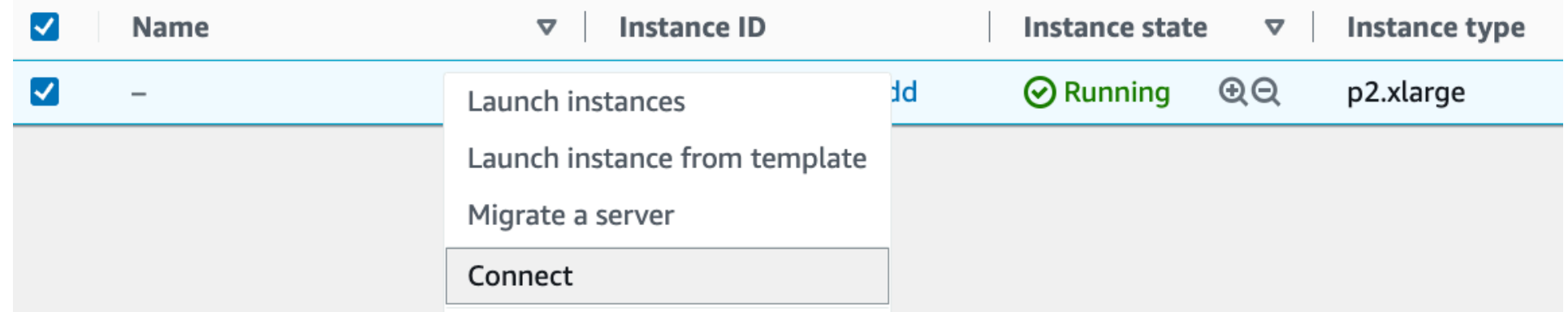

Fig. B.9 View the instance access method.

If this is a new key, it must not be publicly viewable for SSH to work. Go to the folder where you store `D2L_key.pem` and execute the following command to make the key not publicly viewable:

```
chmod 400 D2L_key.pem
```

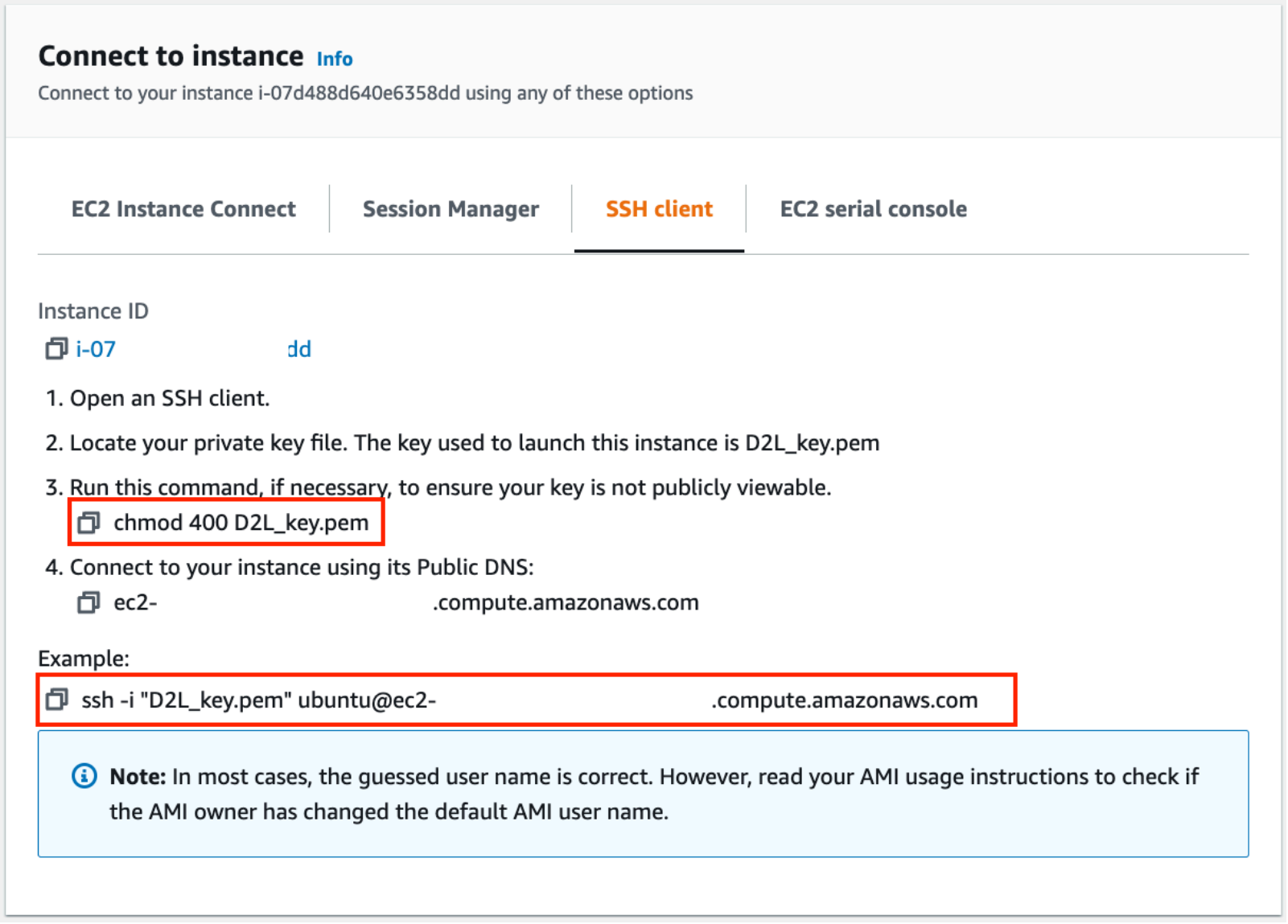

Fig. B.10 View instance access and startup method.

Now, copy the SSH command in the lower red box of Fig. B.10 and paste onto the command line:

```
ssh -i "D2L_key.pem" ubuntu@ec2-xx-xxx-xxx-xxx.y.compute.amazonaws.com
```

When the command line prompts "Are you sure you want to continue connecting (yes/no)", enter "yes" and press Enter to log into the instance.

Your server is ready now.

## B.3.2 Installing CUDA

Before installing CUDA, be sure to update the instance with the latest drivers.

```
sudo apt-get update && sudo apt-get install -y build-essential git libgfortran3
```

Here we download CUDA 12.1. Visit NVIDIA's official repository[297] to find the download link as shown in Fig. B.11.

Copy the instructions and paste them onto the terminal to install CUDA 12.1.

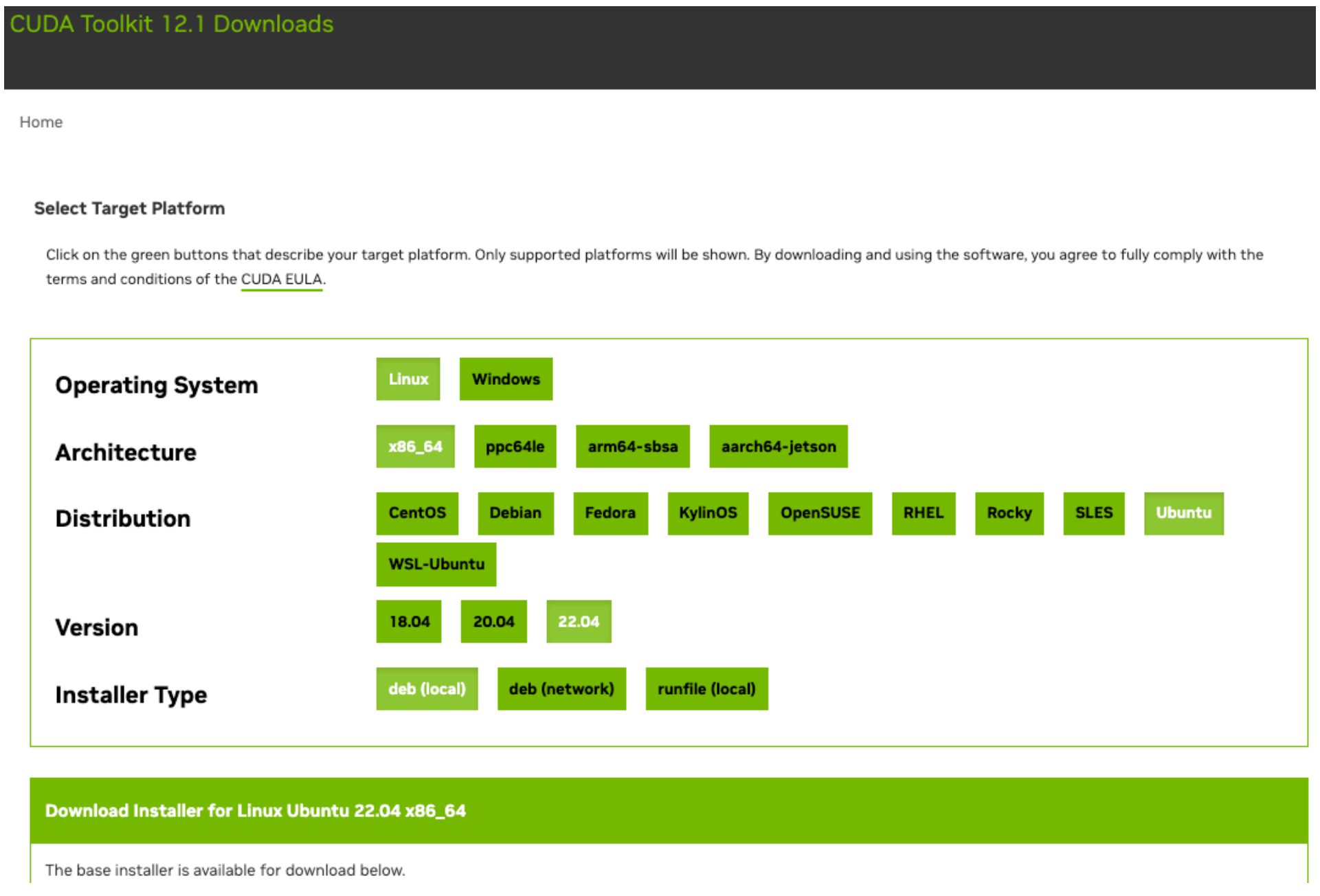

Fig. B.11 Find the CUDA 12.1 download address.

```
# The link and file name are subject to changes
wget https://developer.download.nvidia.com/compute/cuda/repos/ubuntu2204/x86_
↪64/cuda-ubuntu2204.pin
sudo mv cuda-ubuntu2204.pin /etc/apt/preferences.d/cuda-repository-pin-600
wget https://developer.download.nvidia.com/compute/cuda/12.1.0/local_
↪installers/cuda-repo-ubuntu2204-12-1-local_12.1.0-530.30.02-1_amd64.deb
sudo dpkg -i cuda-repo-ubuntu2204-12-1-local_12.1.0-530.30.02-1_amd64.deb
sudo cp /var/cuda-repo-ubuntu2204-12-1-local/cuda-*-keyring.gpg /usr/share/
↪keyrings/
sudo apt-get update
sudo apt-get -y install cuda
```

After installing the program, run the following command to view the GPUs:

```
nvidia-smi
```

Finally, add CUDA to the library path to help other libraries find it, such as appending the following lines to the end of `~/.bashrc`.

```
export PATH="/usr/local/cuda-12.1/bin:$PATH"
export LD_LIBRARY_PATH=${LD_LIBRARY_PATH}:/usr/local/cuda-12.1/lib64
```

### B.3.3 Installing Libraries for Running the Code

To run the code of this book, just follow steps in *Installation* (page xxxiv) for Linux users on the EC2 instance and use the following tips for working on a remote Linux server:

- To download the bash script on the Miniconda installation page, right click the download link and select "Copy Link Address", then execute `wget [copied link address]`.
- After running `~/miniconda3/bin/conda init`, you may execute `source ~/.bashrc` instead of closing and reopening your current shell.

### B.3.4 Running the Jupyter Notebook remotely

To run the Jupyter Notebook remotely you need to use SSH port forwarding. After all, the server in the cloud does not have a monitor or keyboard. For this, log into your server from your desktop (or laptop) as follows:

```
# This command must be run in the local command line
ssh -i "/path/to/key.pem" ubuntu@ec2-xx-xxx-xxx-xxx.y.compute.amazonaws.com -L
↪8889:localhost:8888
```

Next, go to the location of the downloaded code of this book on the EC2 instance, then run:

```
conda activate d2l
jupyter notebook
```

Fig. B.12 shows the possible output after you run the Jupyter Notebook. The last row is the URL for port 8888.

```
(d2l)                                        $ jupyter notebook
[I 05:23:30.157 NotebookApp] Writing notebook server cookie secret to /home/ubuntu/.lo
[I 05:23:30.711 NotebookApp] Serving notebooks from local directory: /home/ubuntu/d2l-
[I 05:23:30.711 NotebookApp] Jupyter Notebook 6.4.4 is running at:
[I 05:23:30.711 NotebookApp] http://localhost:8888/?token=7ae1f41cbd5c6ca705c1ad9f5115
[I 05:23:30.711 NotebookApp]  or http://127.0.0.1:8888/?token=7ae1f41cbd5c6ca705c1ad9f
[I 05:23:30.712 NotebookApp] Use Control-C to stop this server and shut down all kerne
[W 05:23:30.717 NotebookApp] No web browser found: could not locate runnable browser.
[C 05:23:30.717 NotebookApp]

    To access the notebook, open this file in a browser:
        file:///home/ubuntu/.local/share/jupyter/runtime/nbserver-1615-open.html
    Or copy and paste one of these URLs:
        http://localhost:8888/?token=7ae1f41cbd5c6ca705c1ad9f5115931bb8929817fd507ba3
```

Fig. B.12 Output after running the Jupyter Notebook. The last row is the URL for port 8888.

Since you used port forwarding to port 8889, copy the last row in the red box of Fig. B.12, replace "8888" with "8889" in the URL, and open it in your local browser.

### B.3.5 Closing Unused Instances

As cloud services are billed by the time of use, you should close instances that are not being used. Note that there are alternatives:

- "Stopping" an instance means that you will be able to start it again. This is akin to switching off the power for your regular server. However, stopped instances will still be billed a small amount for the hard disk space retained.
- "Terminating" an instance will delete all data associated with it. This includes the disk, hence you cannot start it again. Only do this if you know that you will not need it in the future.

If you want to use the instance as a template for many more instances, right-click on the example in Fig. B.9 and select "Image" → "Create" to create an image of the instance. Once this is complete, select "Instance State" → "Terminate" to terminate the instance. The next time you want to use this instance, you can follow the steps in this section to create an instance based on the saved image. The only difference is that, in "1. Choose AMI" shown in Fig. B.4, you must use the "My AMIs" option on the left to select your saved image. The created instance will retain the information stored on the image hard disk. For example, you will not have to reinstall CUDA and other runtime environments.

### B.3.6 Summary

- We can launch and stop instances on demand without having to buy and build our own computer.
- We need to install CUDA before using the GPU-enabled deep learning framework.
- We can use port forwarding to run the Jupyter Notebook on a remote server.

### B.3.7 Exercises

1. The cloud offers convenience, but it does not come cheap. Find out how to launch spot instances[298] to see how to reduce costs.
2. Experiment with different GPU servers. How fast are they?
3. Experiment with multi-GPU servers. How well can you scale things up?

Discussions[299].

# B.4 Using Google Colab

We introduced how to run this book on AWS in Section B.2 and Section B.3. Another option is running this book on Google Colab[300] if you have a Google account.

To run the code of a section on Colab, simply click the `Colab` button as shown in Fig. B.1.

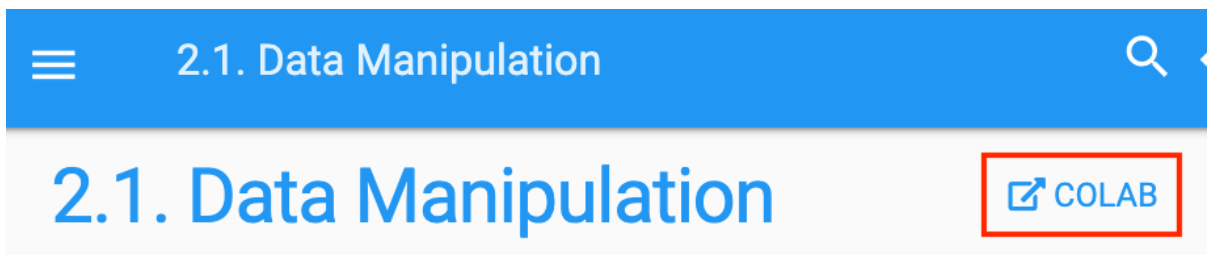

Fig. B.1 Run the code of a section on Colab

If it is your first time to run a code cell, you will receive a warning message as shown in Fig. B.2. Just click "RUN ANYWAY" to ignore it.

Fig. B.2 Ignore the warning message by clicking "RUN ANYWAY".

Next, Colab will connect you to an instance to run the code of this section. Specifically, if a GPU is needed, Colab will be automatically requested for connecting to a GPU instance.

## B.4.1 Summary

- You can use Google Colab to run each section's code in this book.
- Colab will be requested to connect to a GPU instance if a GPU is needed in any section of this book.

## B.4.2 Exercises

1. Open any section of this book using Google Colab.

2. Edit and run any section that requires a GPU using Google Colab.

Discussions[301].

## B.5 Selecting Servers and GPUs

Deep learning training generally requires large amounts of computation. At present GPUs are the most cost-effective hardware accelerators for deep learning. In particular, compared with CPUs, GPUs are cheaper and offer higher performance, often by over an order of magnitude. Furthermore, a single server can support multiple GPUs, up to 8 for high end servers. More typical numbers are up to 4 GPUs for an engineering workstation, since heat, cooling, and power requirements escalate quickly beyond what an office building can support. For larger deployments, cloud computing (e.g., Amazon's P3[302] and G4[303] instances) is a much more practical solution.

### B.5.1 Selecting Servers

There is typically no need to purchase high-end CPUs with many threads since much of the computation occurs on the GPUs. That said, due to the global interpreter lock (GIL) in Python single-thread performance of a CPU can matter in situations where we have 4–8 GPUs. All things equal this suggests that CPUs with a smaller number of cores but a higher clock frequency might be a more economical choice. For example, when choosing between a 6-core 4 GHz and an 8-core 3.5 GHz CPU, the former is much preferable, even though its aggregate speed is less. An important consideration is that GPUs use lots of power and thus dissipate lots of heat. This requires very good cooling and a large enough chassis to use the GPUs. Follow the guidelines below if possible:

1. **Power Supply**. GPUs use significant amounts of power. Budget with up to 350W per device (check for the *peak demand* of the graphics card rather than typical demand, since efficient code can use lots of energy). If your power supply is not up to the demand you will find that your system becomes unstable.

2. **Chassis Size**. GPUs are large and the auxiliary power connectors often need extra space. Also, large chassis are easier to cool.

3. **GPU Cooling**. If you have a large number of GPUs you might want to invest in water cooling. Also, aim for *reference designs* even if they have fewer fans, since they are thin enough to allow for air intake between the devices. If you buy a multi-fan GPU it might be too thick to get enough air when installing multiple GPUs and you will run into thermal throttling.

4. **PCIe Slots**. Moving data to and from the GPU (and exchanging it between GPUs) requires lots of bandwidth. We recommend PCIe 3.0 slots with 16 lanes. If you mount multiple GPUs, be sure to carefully read the motherboard description to ensure that 16× bandwidth is still available when multiple GPUs are used at the same time and that you are getting PCIe 3.0 as opposed to PCIe 2.0 for the additional slots. Some motherboards downgrade to 8× or even 4× bandwidth with multiple GPUs installed. This is partly due to the number of PCIe lanes that the CPU offers.

In short, here are some recommendations for building a deep learning server:

- **Beginner**. Buy a low end GPU with low power consumption (cheap gaming GPUs suitable for deep learning use 150–200W). If you are lucky your current computer supports it.
- **1 GPU**. A low-end CPU with 4 cores will be sufficient and most motherboards suffice. Aim for at least 32 GB DRAM and invest into an SSD for local data access. A power supply with 600W should be sufficient. Buy a GPU with lots of fans.
- **2 GPUs**. A low-end CPU with 4-6 cores will suffice. Aim for 64 GB DRAM and invest into an SSD. You will need in the order of 1000W for two high-end GPUs. In terms of mainboards, make sure that they have *two* PCIe 3.0 x16 slots. If you can, get a mainboard that has two free spaces (60mm spacing) between the PCIe 3.0 x16 slots for extra air. In this case, buy two GPUs with lots of fans.
- **4 GPUs**. Make sure that you buy a CPU with relatively fast single-thread speed (i.e., high clock frequency). You will probably need a CPU with a larger number of PCIe lanes, such as an AMD Threadripper. You will likely need relatively expensive mainboards to get 4 PCIe 3.0 x16 slots since they probably need a PLX to multiplex the PCIe lanes. Buy GPUs with reference design that are narrow and let air in between the GPUs. You need a 1600–2000W power supply and the outlet in your office might not support that. This server will probably run *loud and hot*. You do not want it under your desk. 128 GB of DRAM is recommended. Get an SSD (1–2 TB NVMe) for local storage and a bunch of hard disks in RAID configuration to store your data.
- **8 GPUs**. You need to buy a dedicated multi-GPU server chassis with multiple redundant power supplies (e.g., 2+1 for 1600W per power supply). This will require dual socket server CPUs, 256 GB ECC DRAM, a fast network card (10 GBE recommended), and you will need to check whether the servers support the *physical form factor* of the GPUs. Airflow and wiring placement differ significantly between consumer and server GPUs (e.g., RTX 2080 vs. Tesla V100). This means that you might not be able to install the consumer GPU in a server due to insufficient clearance for the power cable or lack of a suitable wiring harness (as one of the coauthors painfully discovered).

### B.5.2 Selecting GPUs

At present, AMD and NVIDIA are the two main manufacturers of dedicated GPUs. NVIDIA was the first to enter the deep learning field and provides better support for deep learning frameworks via CUDA. Therefore, most buyers choose NVIDIA GPUs.

NVIDIA provides two types of GPUs, targeting individual users (e.g., via the GTX and RTX series) and enterprise users (via its Tesla series). The two types of GPUs provide comparable compute power. However, the enterprise user GPUs generally use (passive) forced cooling, more memory, and ECC (error correcting) memory. These GPUs are more suitable for data centers and usually cost ten times more than consumer GPUs.

If you are a large company with 100+ servers you should consider the NVIDIA Tesla series or alternatively use GPU servers in the cloud. For a lab or a small to medium company with 10+ servers the NVIDIA RTX series is likely most cost effective. You can buy preconfigured servers with Supermicro or Asus chassis that hold 4–8 GPUs efficiently.

GPU vendors typically release a new generation every one to two years, such as the GTX 1000 (Pascal) series released in 2017 and the RTX 2000 (Turing) series released in 2019. Each series offers several different models that provide different performance levels. GPU performance is primarily a combination of the following three parameters:

1. **Compute Power**. Generally we look for 32-bit floating-point compute power. 16-bit floating point training (FP16) is also entering the mainstream. If you are only interested in prediction, you can also use 8-bit integer. The latest generation of Turing GPUs offers 4-bit acceleration. Unfortunately at the time of writing the algorithms for training low-precision networks are not yet widespread.

2. **Memory Size**. As your models become larger or the batches used during training grow bigger, you will need more GPU memory. Check for HBM2 (High Bandwidth Memory) vs. GDDR6 (Graphics DDR) memory. HBM2 is faster but much more expensive.

3. **Memory Bandwidth**. You can only get the most out of your compute power when you have sufficient memory bandwidth. Look for wide memory buses if using GDDR6.

For most users, it is enough to look at compute power. Note that many GPUs offer different types of acceleration. For example, NVIDIA's TensorCores accelerate a subset of operators by 5×. Ensure that your libraries support this. The GPU memory should be no less than 4 GB (8 GB is much better). Try to avoid using the GPU also for displaying a GUI (use the built-in graphics instead). If you cannot avoid it, add an extra 2 GB of RAM for safety.

Fig. B.1 compares the 32-bit floating-point compute power and price of the various GTX 900, GTX 1000 and RTX 2000 series models. The prices suggested are those found on Wikipedia at the time of writing.

We can see a number of things:

1. Within each series, price and performance are roughly proportional. Titan models command a significant premium for the benefit of larger amounts of GPU memory. However, the newer models offer better cost effectiveness, as can be seen by comparing the 980 Ti and 1080 Ti. The price does not appear to improve much for the RTX 2000 series. However, this is due to the fact that they offer far superior low precision performance (FP16, INT8, and INT4).

2. The performance-to-cost ratio of the GTX 1000 series is about two times greater than the 900 series.

3. For the RTX 2000 series the performance (in GFLOPs) is an *affine* function of the price.

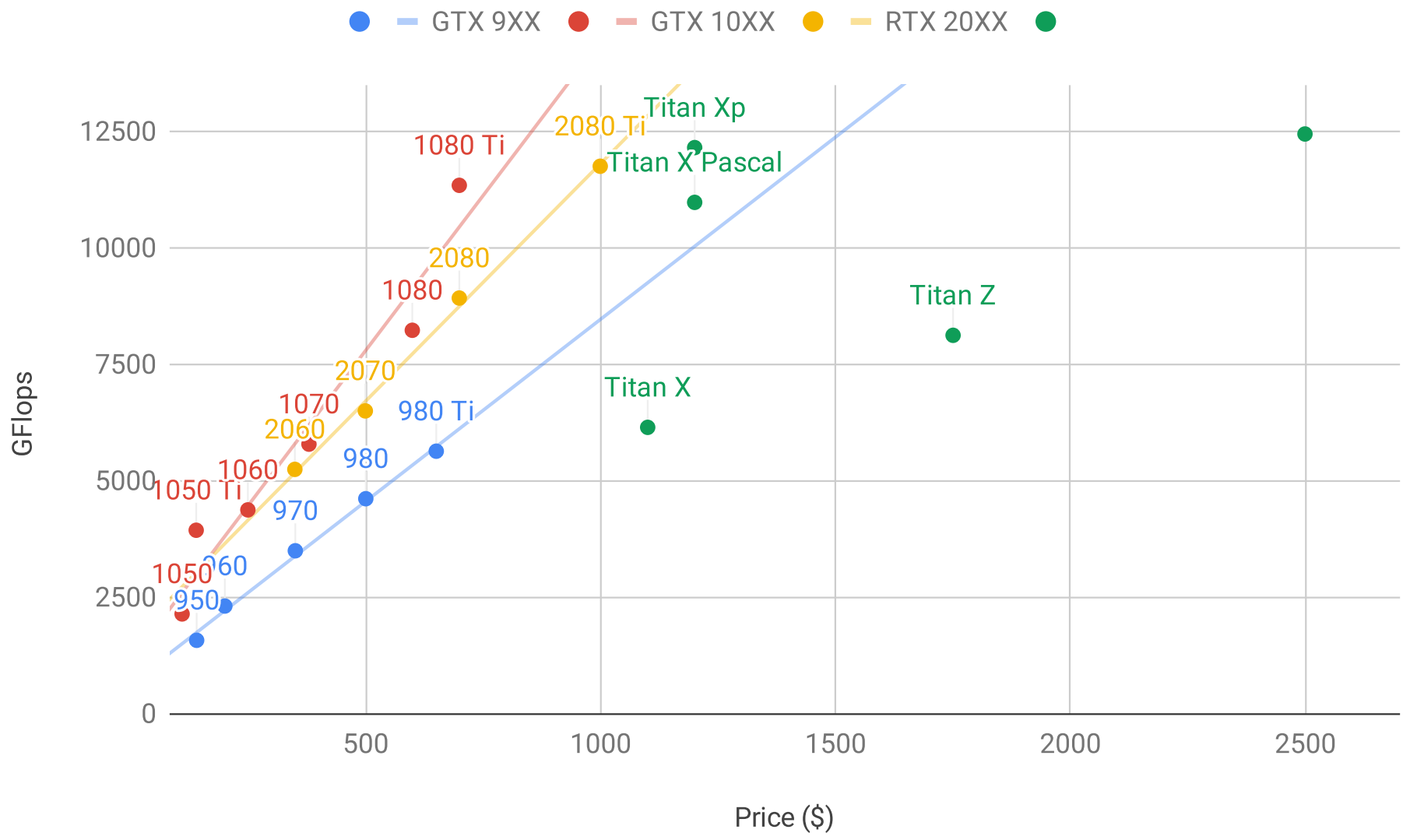

Fig. B.1 Floating-point compute power and price comparison.

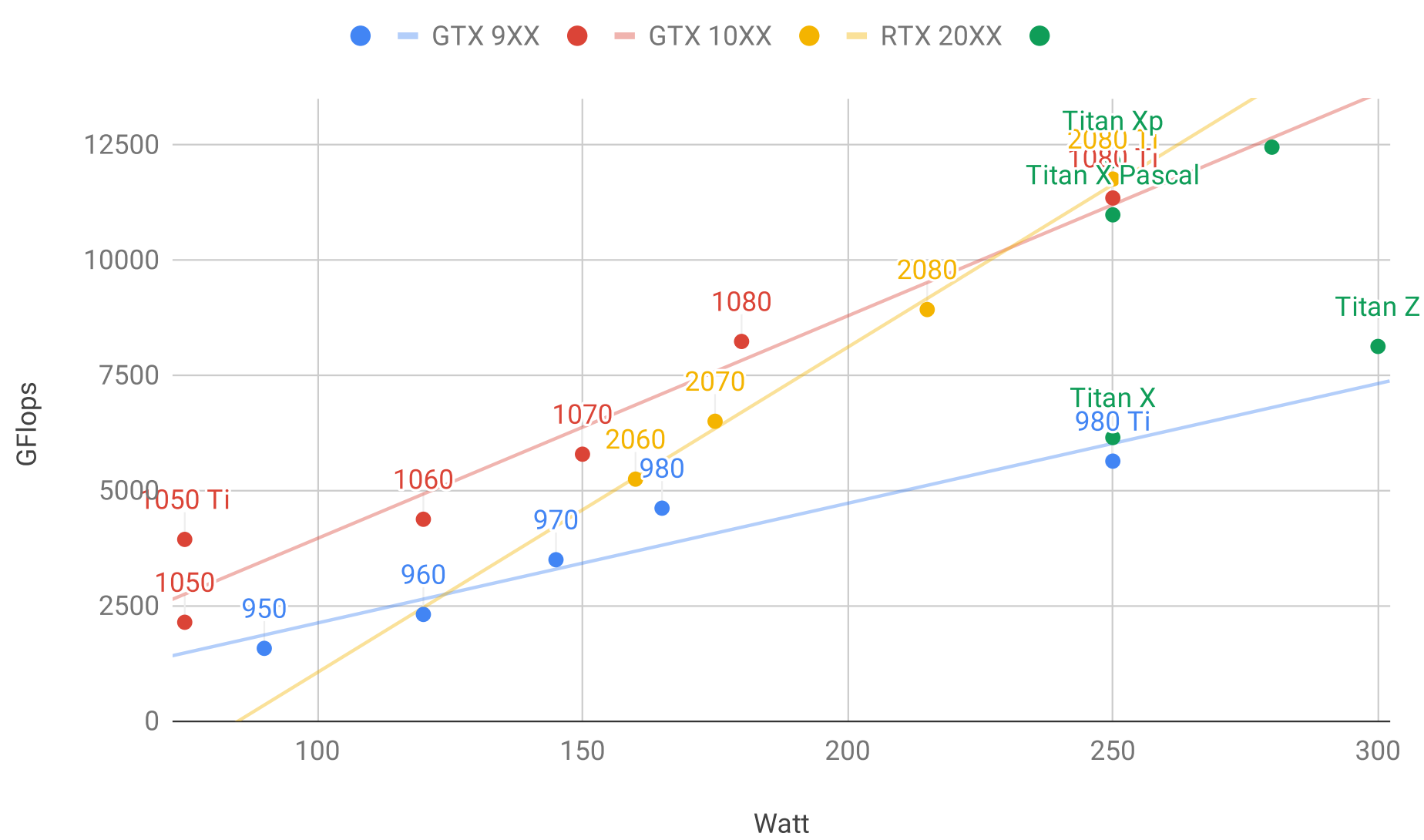

Fig. B.2 Floating-point compute power and energy consumption.

Fig. B.2 shows how energy consumption scales mostly linearly with the amount of computation. Second, later generations are more efficient. This seems to be contradicted by the graph corresponding to the RTX 2000 series. However, this is a consequence of the TensorCores that draw disproportionately much energy.

### B.5.3 Summary

- Watch out for power, PCIe bus lanes, CPU single thread speed, and cooling when building a server.
- You should purchase the latest GPU generation if possible.
- Use the cloud for large deployments.
- High density servers may not be compatible with all GPUs. Check the mechanical and cooling specifications before you buy.
- Use FP16 or lower precision for high efficiency.

Discussions[304].

## B.6 Contributing to This Book

Contributions by readers[305] help us improve this book. If you find a typo, an outdated link, something where you think we missed a citation, where the code does not look elegant or where an explanation is unclear, please contribute back and help us help our readers. While in regular books the delay between print runs (and thus between typo corrections) can be measured in years, it typically takes hours to days to incorporate an improvement in this book. This is all possible due to version control and continuous integration (CI) testing. To do so you need to submit a pull request[306] to the GitHub repository. When your pull request is merged into the code repository by the authors, you will become a contributor.

### B.6.1 Submitting Minor Changes

The most common contributions are editing one sentence or fixing typos. We recommend that you find the source file in the GitHub repository[307] and edit the file directly. For example, you can search the file through the Find file[308] button (Fig. B.1) to locate the source file (a markdown file). Then you click the “Edit this file” button on the upper-right corner to make your changes in the markdown file.

After you are done, fill in your change descriptions in the “Propose file change” panel on the page bottom and then click the “Propose file change” button. It will redirect you to a new page to review your changes (Fig. B.7). If everything is good, you can submit a pull request by clicking the "Create pull request" button.

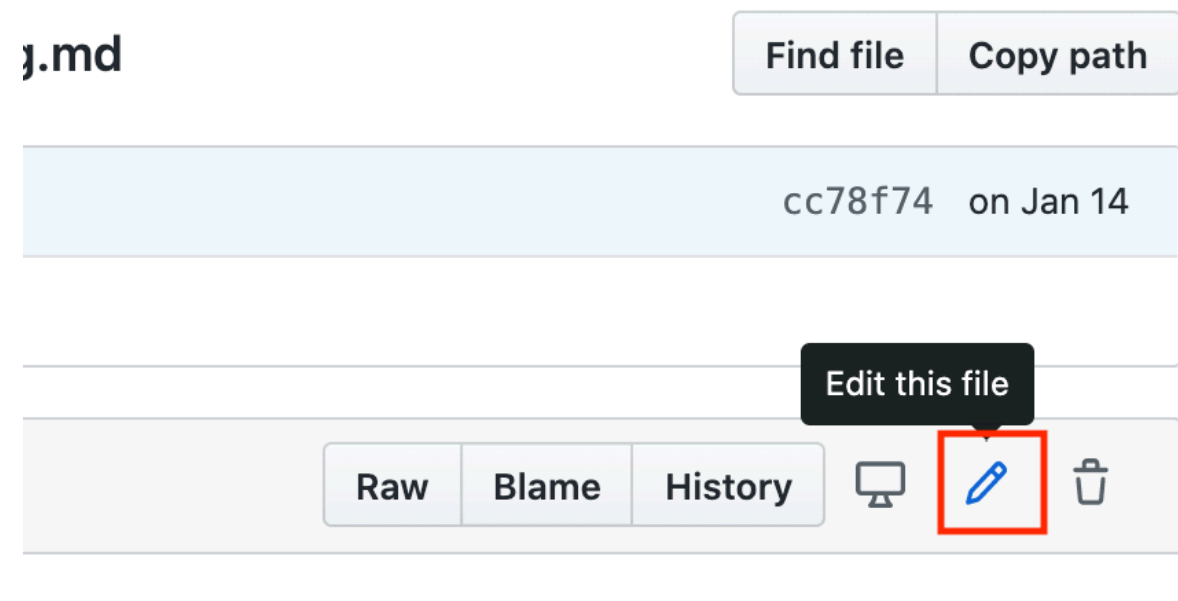

Fig. B.1 Edit the file on Github.

## B.6.2 Proposing Major Changes

If you plan to update a large portion of text or code, then you need to know a little bit more about the format this book is using. The source file is based on the markdown format[309] with a set of extensions through the D2L-Book[310] package such as referring to equations, images, chapters, and citations. You can use any markdown editors to open these files and make your changes.

If you would like to change the code, we recommend that you use the Jupyter Notebook to open these markdown files as described in Section B.1, so that you can run and test your changes. Please remember to clear all outputs before submitting your changes since our CI system will execute the sections you updated to generate outputs.

Some sections may support multiple framework implementations. If you add a new code block, please use `%%tab` to mark this block on the beginning line. For example, `%%tab pytorch` for a PyTorch code block, `%%tab tensorflow` for a TensorFlow code block, or `%%tab all` a shared code block for all implementations. You may refer to the `d2lbook` package for more information.

## B.6.3 Submitting Major Changes

We suggest you to use the standard Git process to submit a major change. In a nutshell the process works as described in Fig. B.2.

We will walk you through the steps in detail. If you are already familiar with Git you can skip this section. For concreteness we assume that the contributor's user name is "astonzhang".

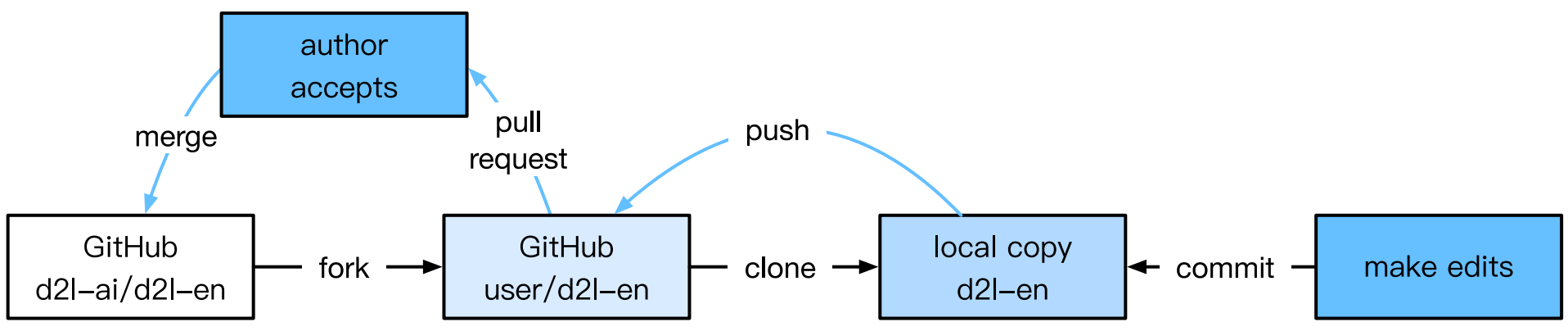

Fig. B.2 Contributing to the book.

## Installing Git

The Git open-source book describes how to install Git[311] . This typically works via `apt install git` on Ubuntu Linux, by installing the Xcode developer tools on macOS, or by using GitHub's desktop client[312] . If you do not have a GitHub account, you need to sign up for one.

## Logging in to GitHub

Enter the address[313] of the book's code repository in your browser. Click on the `Fork` button in the red box at the upper-right of Fig. B.3, to make a copy of the repository of this book. This is now *your copy* and you can change it any way you want.

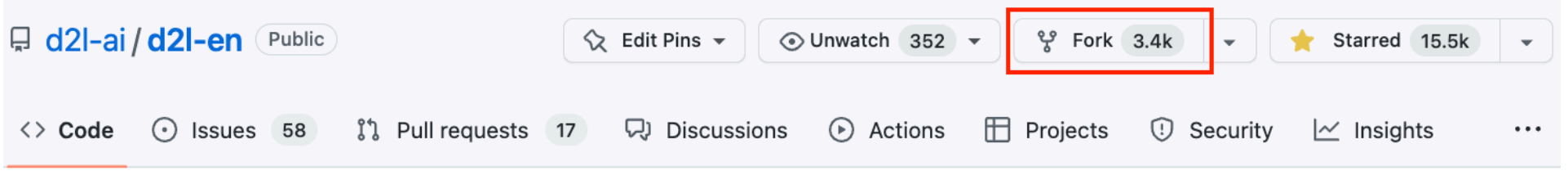

Fig. B.3 The code repository page.

Now, the code repository of this book will be forked (i.e., copied) to your username, such as `astonzhang/d2l-en` shown at the upper-left of Fig. B.4.

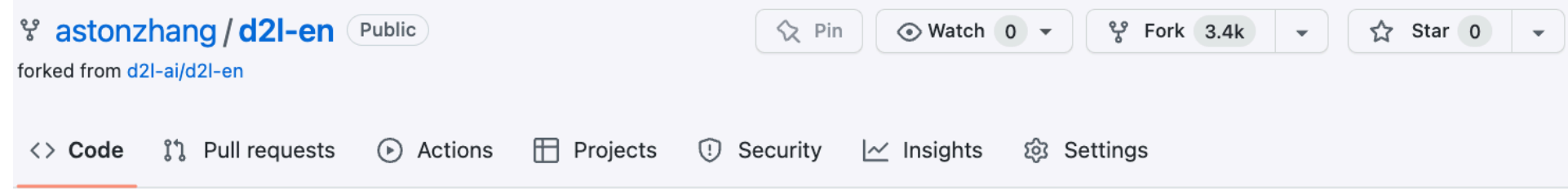

Fig. B.4 The forked code repository.

## Cloning the Repository

To clone the repository (i.e., to make a local copy) we need to get its repository address. The green button in Fig. B.5 displays this. Make sure that your local copy is up to date with the main repository if you decide to keep this fork around for longer. For now simply follow the instructions in *Installation* (page xxxiv) to get started. The main difference is that you are now downloading *your own fork* of the repository.

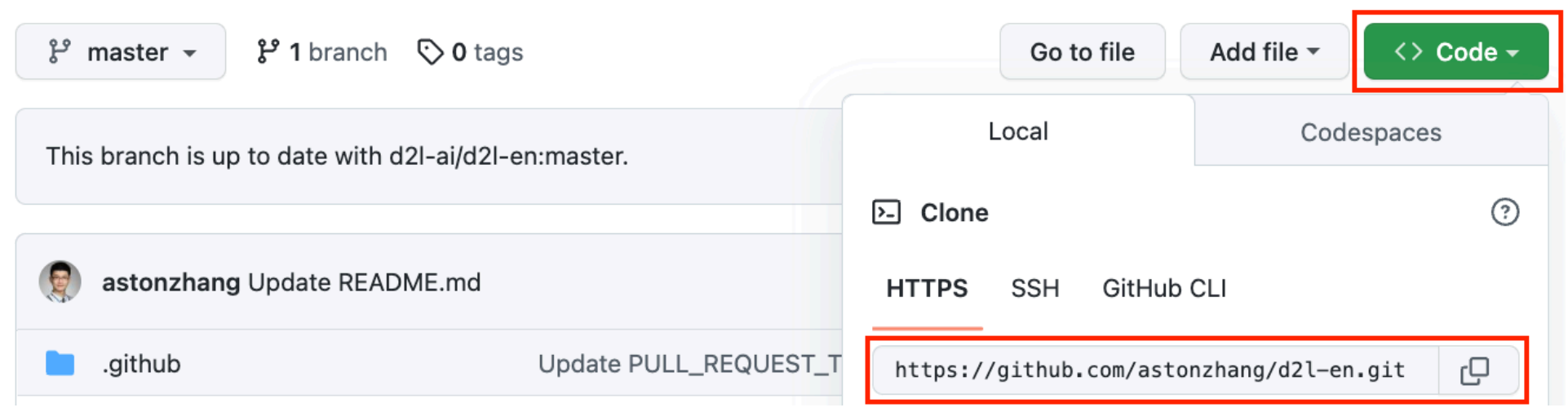

Fig. B.5 Cloning the repository.

```
# Replace your_github_username with your GitHub username
git clone https://github.com/your_github_username/d2l-en.git
```

## Editing and Pushing

Now it is time to edit the book. It is best to edit it in the Jupyter Notebook following instructions in Section B.1. Make the changes and check that they are OK. Assume that we have modified a typo in the file `~/d2l-en/chapter_appendix-tools-for-deep-learning/contributing.md`. You can then check which files you have changed.

At this point Git will prompt that the `chapter_appendix-tools-for-deep-learning/contributing.md` file has been modified.

```
mylaptop:d2l-en me$ git status
On branch master
Your branch is up-to-date with 'origin/master'.

Changes not staged for commit:
  (use "git add <file>..." to update what will be committed)
  (use "git checkout -- <file>..." to discard changes in working directory)

	modified:   chapter_appendix-tools-for-deep-learning/contributing.md
```

After confirming that this is what you want, execute the following command:

```
git add chapter_appendix-tools-for-deep-learning/contributing.md
git commit -m 'Fix a typo in git documentation'
git push
```

The changed code will then be in your personal fork of the repository. To request the addition of your change, you have to create a pull request for the official repository of the book.

### Submitting Pull Requests

As shown in Fig. B.6, go to your fork of the repository on GitHub and select “New pull request”. This will open up a screen that shows you the changes between your edits and what is current in the main repository of the book.

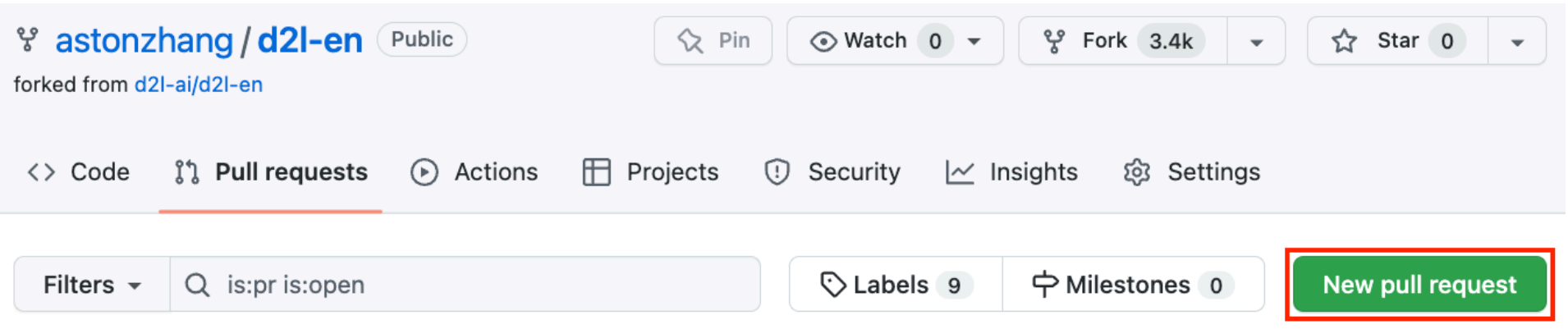

Fig. B.6 New pull request.

Finally, submit a pull request by clicking the button as shown in Fig. B.7. Make sure to describe the changes you have made in the pull request. This will make it easier for the authors to review it and to merge it with the book. Depending on the changes, this might get accepted right away, rejected, or more likely, you will get some feedback on the changes. Once you have incorporated them, you are good to go.

Fig. B.7 Create pull request.

## B.6.4 Summary

- You can use GitHub to contribute to this book.
- You can edit the file on GitHub directly for minor changes.
- For a major change, please fork the repository, edit things locally, and only contribute back once you are ready.
- Pull requests are how contributions are being bundled up. Try not to submit huge pull requests since this makes them hard to understand and incorporate. Better send several smaller ones.

## B.6.5 Exercises

1. Star and fork the `d2l-ai/d2l-en` repository.
2. If you spot anything that needs improvement (e.g., missing a reference), submit a pull request.
3. It is usually a better practice to create a pull request using a new branch. Learn how to do it with Git branching[314].

Discussions[315].

## B.7 Utility Functions and Classes

This section contains the implementations of utility functions and classes used in this book.

```
import collections
import inspect
from IPython import display
from torch import nn
from d2l import torch as d2l
```

Hyperparameters.

```
@d2l.add_to_class(d2l.HyperParameters)  #@save
def save_hyperparameters(self, ignore=[]):
    """Save function arguments into class attributes."""
    frame = inspect.currentframe().f_back
    _, _, _, local_vars = inspect.getargvalues(frame)
    self.hparams = {k:v for k, v in local_vars.items()
                    if k not in set(ignore+['self']) and not k.startswith('_')}
    for k, v in self.hparams.items():
        setattr(self, k, v)
```

Progress bar.

```
@d2l.add_to_class(d2l.ProgressBoard)  #@save
def draw(self, x, y, label, every_n=1):
    Point = collections.namedtuple('Point', ['x', 'y'])
    if not hasattr(self, 'raw_points'):
        self.raw_points = collections.OrderedDict()
        self.data = collections.OrderedDict()
    if label not in self.raw_points:
        self.raw_points[label] = []
        self.data[label] = []
    points = self.raw_points[label]
    line = self.data[label]
    points.append(Point(x, y))
```

```
    if len(points) != every_n:
        return
    mean = lambda x: sum(x) / len(x)
    line.append(Point(mean([p.x for p in points]),
                      mean([p.y for p in points])))
    points.clear()
    if not self.display:
        return
    d2l.use_svg_display()
    if self.fig is None:
        self.fig = d2l.plt.figure(figsize=self.figsize)
    plt_lines, labels = [], []
    for (k, v), ls, color in zip(self.data.items(), self.ls, self.colors):
        plt_lines.append(d2l.plt.plot([p.x for p in v], [p.y for p in v],
                                      linestyle=ls, color=color)[0])
        labels.append(k)
    axes = self.axes if self.axes else d2l.plt.gca()
    if self.xlim: axes.set_xlim(self.xlim)
    if self.ylim: axes.set_ylim(self.ylim)
    if not self.xlabel: self.xlabel = self.x
    axes.set_xlabel(self.xlabel)
    axes.set_ylabel(self.ylabel)
    axes.set_xscale(self.xscale)
    axes.set_yscale(self.yscale)
    axes.legend(plt_lines, labels)
    display.display(self.fig)
    display.clear_output(wait=True)
```

Add FrozenLake enviroment

```
def frozen_lake(seed): #@save
    # See https://www.gymlibrary.dev/environments/toy_text/frozen_lake/ to␣
↪learn more about this env
    # How to process env.P.items is adpated from https://sites.google.com/view/
↪deep-rl-bootcamp/labs
    import gym

    env = gym.make('FrozenLake-v1', is_slippery=False)
    env.seed(seed)
    env.action_space.np_random.seed(seed)
    env.action_space.seed(seed)
    env_info = {}
    env_info['desc'] = env.desc  # 2D array specifying what each grid item␣
↪means
    env_info['num_states'] = env.nS  # Number of observations/states or obs/
↪state dim
    env_info['num_actions'] = env.nA  # Number of actions or action dim
    # Define indices for (transition probability, nextstate, reward, done)␣
↪tuple
    env_info['trans_prob_idx'] = 0  # Index of transition probability entry
    env_info['nextstate_idx'] = 1  # Index of next state entry
    env_info['reward_idx'] = 2  # Index of reward entry
    env_info['done_idx'] = 3  # Index of done entry
    env_info['mdp'] = {}
```

```
    env_info['env'] = env

    for (s, others) in env.P.items():
        # others(s) = {a0: [ (p(s'|s,a0), s', reward, done),...], a1:[...], ...
↪}

        for (a, pxrds) in others.items():
            # pxrds is [(p1,next1,r1,d1),(p2,next2,r2,d2),..].
            # e.g. [(0.3, 0, 0, False), (0.3, 0, 0, False), (0.3, 4, 1, False)]
            env_info['mdp'][(s,a)] = pxrds

    return env_info
```

Create enviroment

```
def make_env(name ='', seed=0): #@save
    # Input parameters:
    # name: specifies a gym environment.
    # For Value iteration, only FrozenLake-v1 is supported.
    if name == 'FrozenLake-v1':
        return frozen_lake(seed)

    else:
        raise ValueError("%s env is not supported in this Notebook")
```

Show value function

```
def show_value_function_progress(env_desc, V, pi): #@save
    # This function visualizes how value and policy changes over time.
    # V: [num_iters, num_states]
    # pi: [num_iters, num_states]
    # How to visualize value function is adapted (but changed) from: https://
↪sites.google.com/view/deep-rl-bootcamp/labs

    num_iters = V.shape[0]
    fig, ax  = plt.subplots(figsize=(15, 15))

    for k in range(V.shape[0]):
        plt.subplot(4, 4, k + 1)
        plt.imshow(V[k].reshape(4,4), cmap="bone")
        ax = plt.gca()
        ax.set_xticks(np.arange(0, 5)-.5, minor=True)
        ax.set_yticks(np.arange(0, 5)-.5, minor=True)
        ax.grid(which="minor", color="w", linestyle='-', linewidth=3)
        ax.tick_params(which="minor", bottom=False, left=False)
        ax.set_xticks([])
        ax.set_yticks([])

        # LEFT action: 0, DOWN action: 1
        # RIGHT action: 2, UP action: 3
        action2dxdy = {0:(-.25, 0),1: (0, .25),
                       2:(0.25, 0),3: (-.25, 0)}
```

```
        for y in range(4):
            for x in range(4):
                action = pi[k].reshape(4,4)[y, x]
                dx, dy = action2dxdy[action]

                if env_desc[y,x].decode() == 'H':
                    ax.text(x, y, str(env_desc[y,x].decode()),
                       ha="center", va="center", color="y",
                         size=20, fontweight='bold')

                elif env_desc[y,x].decode() == 'G':
                    ax.text(x, y, str(env_desc[y,x].decode()),
                       ha="center", va="center", color="w",
                         size=20, fontweight='bold')

                else:
                    ax.text(x, y, str(env_desc[y,x].decode()),
                       ha="center", va="center", color="g",
                         size=15, fontweight='bold')

                # No arrow for cells with G and H labels
                if env_desc[y,x].decode() != 'G' and env_desc[y,x].decode() !=
↪'H':
                    ax.arrow(x, y, dx, dy, color='r', head_width=0.2, head_
↪length=0.15)

        ax.set_title("Step = "  + str(k + 1), fontsize=20)

    fig.tight_layout()
    plt.show()
```

Show Q function

```
def show_Q_function_progress(env_desc, V_all, pi_all): #@save
    # This function visualizes how value and policy changes over time.
    # V: [num_iters, num_states]
    # pi: [num_iters, num_states]

    # We want to only shows few values
    num_iters_all = V_all.shape[0]
    num_iters = num_iters_all // 10

    vis_indx = np.arange(0, num_iters_all, num_iters).tolist()
    vis_indx.append(num_iters_all - 1)
    V = np.zeros((len(vis_indx), V_all.shape[1]))
    pi = np.zeros((len(vis_indx), V_all.shape[1]))

    for c, i in enumerate(vis_indx):
        V[c]  = V_all[i]
        pi[c] = pi_all[i]

    num_iters = V.shape[0]
    fig, ax = plt.subplots(figsize=(15, 15))
```

```
    for k in range(V.shape[0]):
        plt.subplot(4, 4, k + 1)
        plt.imshow(V[k].reshape(4,4), cmap="bone")
        ax = plt.gca()
        ax.set_xticks(np.arange(0, 5)-.5, minor=True)
        ax.set_yticks(np.arange(0, 5)-.5, minor=True)
        ax.grid(which="minor", color="w", linestyle='-', linewidth=3)
        ax.tick_params(which="minor", bottom=False, left=False)
        ax.set_xticks([])
        ax.set_yticks([])

        # LEFT action: 0, DOWN action: 1
        # RIGHT action: 2, UP action: 3
        action2dxdy = {0:(-.25, 0),1:(0, .25),
                       2:(0.25, 0),3:(-.25, 0)}

        for y in range(4):
            for x in range(4):
                action = pi[k].reshape(4,4)[y, x]
                dx, dy = action2dxdy[action]

                if env_desc[y,x].decode() == 'H':
                    ax.text(x, y, str(env_desc[y,x].decode()),
                       ha="center", va="center", color="y",
                         size=20, fontweight='bold')

                elif env_desc[y,x].decode() == 'G':
                    ax.text(x, y, str(env_desc[y,x].decode()),
                       ha="center", va="center", color="w",
                         size=20, fontweight='bold')

                else:
                    ax.text(x, y, str(env_desc[y,x].decode()),
                       ha="center", va="center", color="g",
                         size=15, fontweight='bold')

                # No arrow for cells with G and H labels
                if env_desc[y,x].decode() != 'G' and env_desc[y,x].decode() !=
↪'H':
                    ax.arrow(x, y, dx, dy, color='r', head_width=0.2, head_
↪length=0.15)

        ax.set_title("Step = "  + str(vis_indx[k] + 1), fontsize=20)

    fig.tight_layout()
    plt.show()
```

Trainer

A bunch of functions that will be deprecated:

```
def load_array(data_arrays, batch_size, is_train=True):  #@save
    """Construct a PyTorch data iterator."""
    dataset = torch.utils.data.TensorDataset(*data_arrays)
```

```
    return torch.utils.data.DataLoader(dataset, batch_size, shuffle=is_train)

def synthetic_data(w, b, num_examples):  #@save
    """Generate y = Xw + b + noise."""
    X = torch.normal(0, 1, (num_examples, len(w)))
    y = torch.matmul(X, w) + b
    y += torch.normal(0, 0.01, y.shape)
    return X, y.reshape((-1, 1))

def sgd(params, lr, batch_size): #@save
    """Minibatch stochastic gradient descent."""
    with torch.no_grad():
        for param in params:
            param -= lr * param.grad / batch_size
            param.grad.zero_()

def get_dataloader_workers():  #@save
    """Use 4 processes to read the data."""
    return 4

def load_data_fashion_mnist(batch_size, resize=None):  #@save
    """Download the Fashion-MNIST dataset and then load it into memory."""
    trans = [transforms.ToTensor()]
    if resize:
        trans.insert(0, transforms.Resize(resize))
    trans = transforms.Compose(trans)
    mnist_train = torchvision.datasets.FashionMNIST(
        root="../data", train=True, transform=trans, download=True)
    mnist_test = torchvision.datasets.FashionMNIST(
        root="../data", train=False, transform=trans, download=True)
    return (torch.utils.data.DataLoader(mnist_train, batch_size, shuffle=True,
                                        num_workers=get_dataloader_workers()),
            torch.utils.data.DataLoader(mnist_test, batch_size, shuffle=False,
                                        num_workers=get_dataloader_workers()))

def evaluate_accuracy_gpu(net, data_iter, device=None): #@save
    """Compute the accuracy for a model on a dataset using a GPU."""
    if isinstance(net, nn.Module):
        net.eval()  # Set the model to evaluation mode
        if not device:
            device = next(iter(net.parameters())).device
    # No. of correct predictions, no. of predictions
    metric = d2l.Accumulator(2)

    with torch.no_grad():
        for X, y in data_iter:
            if isinstance(X, list):
                # Required for BERT Fine-tuning (to be covered later)
                X = [x.to(device) for x in X]
            else:
                X = X.to(device)
            y = y.to(device)
            metric.add(d2l.accuracy(net(X), y), y.numel())
    return metric[0] / metric[1]
```

```
#@save
def train_ch6(net, train_iter, test_iter, num_epochs, lr, device):
    """Train a model with a GPU (defined in Chapter 6)."""
    def init_weights(m):
        if type(m) == nn.Linear or type(m) == nn.Conv2d:
            nn.init.xavier_uniform_(m.weight)
    net.apply(init_weights)
    print('training on', device)
    net.to(device)
    optimizer = torch.optim.SGD(net.parameters(), lr=lr)
    loss = nn.CrossEntropyLoss()
    animator = d2l.Animator(xlabel='epoch', xlim=[1, num_epochs],
                            legend=['train loss', 'train acc', 'test acc'])
    timer, num_batches = d2l.Timer(), len(train_iter)
    for epoch in range(num_epochs):
        # Sum of training loss, sum of training accuracy, no. of examples
        metric = d2l.Accumulator(3)
        net.train()
        for i, (X, y) in enumerate(train_iter):
            timer.start()
            optimizer.zero_grad()
            X, y = X.to(device), y.to(device)
            y_hat = net(X)
            l = loss(y_hat, y)
            l.backward()
            optimizer.step()
            with torch.no_grad():
                metric.add(l * X.shape[0], d2l.accuracy(y_hat, y), X.shape[0])
            timer.stop()
            train_l = metric[0] / metric[2]
            train_acc = metric[1] / metric[2]
            if (i + 1) % (num_batches // 5) == 0 or i == num_batches - 1:
                animator.add(epoch + (i + 1) / num_batches,
                             (train_l, train_acc, None))
        test_acc = evaluate_accuracy_gpu(net, test_iter)
        animator.add(epoch + 1, (None, None, test_acc))
    print(f'loss {train_l:.3f}, train acc {train_acc:.3f}, '
          f'test acc {test_acc:.3f}')
    print(f'{metric[2] * num_epochs / timer.sum():.1f} examples/sec '
          f'on {str(device)}')
```

```
def show_images(imgs, num_rows, num_cols, titles=None, scale=1.5):  #@save
    """Plot a list of images."""
    figsize = (num_cols * scale, num_rows * scale)
    _, axes = d2l.plt.subplots(num_rows, num_cols, figsize=figsize)
    axes = axes.flatten()
    for i, (ax, img) in enumerate(zip(axes, imgs)):
        try:
            img = img.detach().numpy()
        except:
            pass
        ax.imshow(img)
        ax.axes.get_xaxis().set_visible(False)
```

```
        ax.axes.get_yaxis().set_visible(False)
        if titles:
            ax.set_title(titles[i])
    return axes
```

```
def linreg(X, w, b):  #@save
    """The linear regression model."""
    return torch.matmul(X, w) + b

def squared_loss(y_hat, y):  #@save
    """Squared loss."""
    return (y_hat - y.reshape(y_hat.shape)) ** 2 / 2

def get_fashion_mnist_labels(labels):  #@save
    """Return text labels for the Fashion-MNIST dataset."""
    text_labels = ['t-shirt', 'trouser', 'pullover', 'dress', 'coat',
                   'sandal', 'shirt', 'sneaker', 'bag', 'ankle boot']
    return [text_labels[int(i)] for i in labels]

class Animator:  #@save
    """For plotting data in animation."""
    def __init__(self, xlabel=None, ylabel=None, legend=None, xlim=None,
                 ylim=None, xscale='linear', yscale='linear',
                 fmts=('-', 'm--', 'g-.', 'r:'), nrows=1, ncols=1,
                 figsize=(3.5, 2.5)):
        # Incrementally plot multiple lines
        if legend is None:
            legend = []
        d2l.use_svg_display()
        self.fig, self.axes = d2l.plt.subplots(nrows, ncols, figsize=figsize)
        if nrows * ncols == 1:
            self.axes = [self.axes, ]
        # Use a lambda function to capture arguments
        self.config_axes = lambda: d2l.set_axes(
            self.axes[0], xlabel, ylabel, xlim, ylim, xscale, yscale, legend)
        self.X, self.Y, self.fmts = None, None, fmts

    def add(self, x, y):
        # Add multiple data points into the figure
        if not hasattr(y, "__len__"):
            y = [y]
        n = len(y)
        if not hasattr(x, "__len__"):
            x = [x] * n
        if not self.X:
            self.X = [[] for _ in range(n)]
        if not self.Y:
            self.Y = [[] for _ in range(n)]
        for i, (a, b) in enumerate(zip(x, y)):
            if a is not None and b is not None:
                self.X[i].append(a)
                self.Y[i].append(b)
        self.axes[0].cla()
        for x, y, fmt in zip(self.X, self.Y, self.fmts):
```

```
            self.axes[0].plot(x, y, fmt)
        self.config_axes()
        display.display(self.fig)
        display.clear_output(wait=True)

class Accumulator:  #@save
    """For accumulating sums over `n` variables."""
    def __init__(self, n):
        self.data = [0.0] * n

    def add(self, *args):
        self.data = [a + float(b) for a, b in zip(self.data, args)]

    def reset(self):
        self.data = [0.0] * len(self.data)

    def __getitem__(self, idx):
        return self.data[idx]

def accuracy(y_hat, y):  #@save
    """Compute the number of correct predictions."""
    if len(y_hat.shape) > 1 and y_hat.shape[1] > 1:
        y_hat = y_hat.argmax(axis=1)
    cmp = y_hat.type(y.dtype) == y
    return float(cmp.type(y.dtype).sum())
```

```
import hashlib
import os
import tarfile
import zipfile
import requests

def download(url, folder='../data', sha1_hash=None):  #@save
    """Download a file to folder and return the local filepath."""
    if not url.startswith('http'):
        # For back compatability
        url, sha1_hash = DATA_HUB[url]
    os.makedirs(folder, exist_ok=True)
    fname = os.path.join(folder, url.split('/')[-1])
    # Check if hit cache
    if os.path.exists(fname) and sha1_hash:
        sha1 = hashlib.sha1()
        with open(fname, 'rb') as f:
            while True:
                data = f.read(1048576)
                if not data:
                    break
                sha1.update(data)
        if sha1.hexdigest() == sha1_hash:
            return fname
    # Download
    print(f'Downloading {fname} from {url}...')
```

```
    r = requests.get(url, stream=True, verify=True)
    with open(fname, 'wb') as f:
        f.write(r.content)
    return fname

def extract(filename, folder=None):  #@save
    """Extract a zip/tar file into folder."""
    base_dir = os.path.dirname(filename)
    _, ext = os.path.splitext(filename)
    assert ext in ('.zip', '.tar', '.gz'), 'Only support zip/tar files.'
    if ext == '.zip':
        fp = zipfile.ZipFile(filename, 'r')
    else:
        fp = tarfile.open(filename, 'r')
    if folder is None:
        folder = base_dir
    fp.extractall(folder)
```

```
def download_extract(name, folder=None):  #@save
    """Download and extract a zip/tar file."""
    fname = download(name)
    base_dir = os.path.dirname(fname)
    data_dir, ext = os.path.splitext(fname)
    if ext == '.zip':
        fp = zipfile.ZipFile(fname, 'r')
    elif ext in ('.tar', '.gz'):
        fp = tarfile.open(fname, 'r')
    else:
        assert False, 'Only zip/tar files can be extracted.'
    fp.extractall(base_dir)
    return os.path.join(base_dir, folder) if folder else data_dir

def tokenize(lines, token='word'):  #@save
    """Split text lines into word or character tokens."""
    assert token in ('word', 'char'), 'Unknown token type: ' + token
    return [line.split() if token == 'word' else list(line) for line in lines]
```

```
def evaluate_loss(net, data_iter, loss):  #@save
    """Evaluate the loss of a model on the given dataset."""
    metric = d2l.Accumulator(2)  # Sum of losses, no. of examples
    for X, y in data_iter:
        out = net(X)
        y = y.reshape(out.shape)
        l = loss(out, y)
        metric.add(l.sum(), l.numel())
    return metric[0] / metric[1]
```

```
def grad_clipping(net, theta):  #@save
    """Clip the gradient."""
    if isinstance(net, nn.Module):
        params = [p for p in net.parameters() if p.requires_grad]
```

```
    else:
        params = net.params
    norm = torch.sqrt(sum(torch.sum((p.grad ** 2)) for p in params))
    if norm > theta:
        for param in params:
            param.grad[:] *= theta / norm
```

More for the attention chapter.

```
#@save
d2l.DATA_HUB['fra-eng'] = (d2l.DATA_URL + 'fra-eng.zip',
                           '94646ad1522d915e7b0f9296181140edcf86a4f5')

#@save
def read_data_nmt():
    """Load the English-French dataset."""
    data_dir = d2l.download_extract('fra-eng')
    with open(os.path.join(data_dir, 'fra.txt'), 'r', encoding='utf-8') as f:
        return f.read()

#@save
def preprocess_nmt(text):
    """Preprocess the English-French dataset."""
    def no_space(char, prev_char):
        return char in set(',.!?') and prev_char != ' '

    # Replace non-breaking space with space, and convert uppercase letters to
    # lowercase ones
    text = text.replace('\u202f', ' ').replace('\xa0', ' ').lower()
    # Insert space between words and punctuation marks
    out = [' ' + char if i > 0 and no_space(char, text[i - 1]) else char
           for i, char in enumerate(text)]
    return ''.join(out)

#@save
def tokenize_nmt(text, num_examples=None):
    """Tokenize the English-French dataset."""
    source, target = [], []
    for i, line in enumerate(text.split('\n')):
        if num_examples and i > num_examples:
            break
        parts = line.split('\t')
        if len(parts) == 2:
            source.append(parts[0].split(' '))
            target.append(parts[1].split(' '))
    return source, target

#@save
def truncate_pad(line, num_steps, padding_token):
    """Truncate or pad sequences."""
    if len(line) > num_steps:
        return line[:num_steps]  # Truncate
    return line + [padding_token] * (num_steps - len(line))  # Pad
```

```
#@save
def build_array_nmt(lines, vocab, num_steps):
    """Transform text sequences of machine translation into minibatches."""
    lines = [vocab[l] for l in lines]
    lines = [l + [vocab['<eos>']] for l in lines]
    array = torch.tensor([truncate_pad(
        l, num_steps, vocab['<pad>']) for l in lines])
    valid_len = (array != vocab['<pad>']).type(torch.int32).sum(1)
    return array, valid_len

#@save
def load_data_nmt(batch_size, num_steps, num_examples=600):
    """Return the iterator and the vocabularies of the translation dataset."""
    text = preprocess_nmt(read_data_nmt())
    source, target = tokenize_nmt(text, num_examples)
    src_vocab = d2l.Vocab(source, min_freq=2,
                          reserved_tokens=['<pad>', '<bos>', '<eos>'])
    tgt_vocab = d2l.Vocab(target, min_freq=2,
                          reserved_tokens=['<pad>', '<bos>', '<eos>'])
    src_array, src_valid_len = build_array_nmt(source, src_vocab, num_steps)
    tgt_array, tgt_valid_len = build_array_nmt(target, tgt_vocab, num_steps)
    data_arrays = (src_array, src_valid_len, tgt_array, tgt_valid_len)
    data_iter = d2l.load_array(data_arrays, batch_size)
    return data_iter, src_vocab, tgt_vocab
```

```
#@save
def sequence_mask(X, valid_len, value=0):
    """Mask irrelevant entries in sequences."""
    maxlen = X.size(1)
    mask = torch.arange((maxlen), dtype=torch.float32,
                        device=X.device)[None, :] < valid_len[:, None]
    X[~mask] = value
    return X

#@save
class MaskedSoftmaxCELoss(nn.CrossEntropyLoss):
    """The softmax cross-entropy loss with masks."""
    # `pred` shape: (`batch_size`, `num_steps`, `vocab_size`)
    # `label` shape: (`batch_size`, `num_steps`)
    # `valid_len` shape: (`batch_size`,)
    def forward(self, pred, label, valid_len):
        weights = torch.ones_like(label)
        weights = sequence_mask(weights, valid_len)
        self.reduction='none'
        unweighted_loss = super(MaskedSoftmaxCELoss, self).forward(
            pred.permute(0, 2, 1), label)
        weighted_loss = (unweighted_loss * weights).mean(dim=1)
        return weighted_loss

#@save
```

```
def train_seq2seq(net, data_iter, lr, num_epochs, tgt_vocab, device):
    """Train a model for sequence to sequence."""
    def xavier_init_weights(m):
        if type(m) == nn.Linear:
            nn.init.xavier_uniform_(m.weight)
        if type(m) == nn.GRU:
            for param in m._flat_weights_names:
                if "weight" in param:
                    nn.init.xavier_uniform_(m._parameters[param])
    net.apply(xavier_init_weights)
    net.to(device)
    optimizer = torch.optim.Adam(net.parameters(), lr=lr)
    loss = MaskedSoftmaxCELoss()
    net.train()
    animator = d2l.Animator(xlabel='epoch', ylabel='loss',
                            xlim=[10, num_epochs])
    for epoch in range(num_epochs):
        timer = d2l.Timer()
        metric = d2l.Accumulator(2)  # Sum of training loss, no. of tokens
        for batch in data_iter:
            optimizer.zero_grad()
            X, X_valid_len, Y, Y_valid_len = [x.to(device) for x in batch]
            bos = torch.tensor([tgt_vocab['<bos>']] * Y.shape[0],
                               device=device).reshape(-1, 1)
            dec_input = torch.cat([bos, Y[:, :-1]], 1)  # Teacher forcing
            Y_hat, _ = net(X, dec_input, X_valid_len)
            l = loss(Y_hat, Y, Y_valid_len)
            l.sum().backward()  # Make the loss scalar for `backward`
            d2l.grad_clipping(net, 1)
            num_tokens = Y_valid_len.sum()
            optimizer.step()
            with torch.no_grad():
                metric.add(l.sum(), num_tokens)
        if (epoch + 1) % 10 == 0:
            animator.add(epoch + 1, (metric[0] / metric[1],))
    print(f'loss {metric[0] / metric[1]:.3f}, {metric[1] / timer.stop():.1f} '
          f'tokens/sec on {str(device)}')

#@save
def predict_seq2seq(net, src_sentence, src_vocab, tgt_vocab, num_steps,
                    device, save_attention_weights=False):
    """Predict for sequence to sequence."""
    # Set `net` to eval mode for inference
    net.eval()
    src_tokens = src_vocab[src_sentence.lower().split(' ')] + [
        src_vocab['<eos>']]
    enc_valid_len = torch.tensor([len(src_tokens)], device=device)
    src_tokens = d2l.truncate_pad(src_tokens, num_steps, src_vocab['<pad>'])
    # Add the batch axis
    enc_X = torch.unsqueeze(
        torch.tensor(src_tokens, dtype=torch.long, device=device), dim=0)
    enc_outputs = net.encoder(enc_X, enc_valid_len)
    dec_state = net.decoder.init_state(enc_outputs, enc_valid_len)
    # Add the batch axis
```

```
    dec_X = torch.unsqueeze(torch.tensor(
        [tgt_vocab['<bos>']], dtype=torch.long, device=device), dim=0)
    output_seq, attention_weight_seq = [], []
    for _ in range(num_steps):
        Y, dec_state = net.decoder(dec_X, dec_state)
        # We use the token with the highest prediction likelihood as input
        # of the decoder at the next time step
        dec_X = Y.argmax(dim=2)
        pred = dec_X.squeeze(dim=0).type(torch.int32).item()
        # Save attention weights (to be covered later)
        if save_attention_weights:
            attention_weight_seq.append(net.decoder.attention_weights)
        # Once the end-of-sequence token is predicted, the generation of the
        # output sequence is complete
        if pred == tgt_vocab['<eos>']:
            break
        output_seq.append(pred)
    return ' '.join(tgt_vocab.to_tokens(output_seq)), attention_weight_seq
```

## B.8 The `d2l` API Document

This section displays classes and functions (sorted alphabetically) in the `d2l` package, showing where they are defined in the book so you can find more detailed implementations and explanations. See also the source code on the GitHub repository[316].

### B.8.1 Classes

`class d2l.torch.AdditiveAttention(`*num_hiddens*, *dropout*, ***kwargs*`)`

Bases: `Module`

Additive attention.

Defined in Section 11.3.2

`forward(`*queries*, *keys*, *values*, *valid_lens*`)`

Defines the computation performed at every call.

Should be overridden by all subclasses.

**Note:** Although the recipe for forward pass needs to be defined within this function, one should call the `Module` (page 1113) instance afterwards instead of this since the former takes care of running the registered hooks while the latter silently ignores them.

`training: bool`

`class d2l.torch.AddNorm(`*norm_shape*, *dropout*`)`

Bases: `Module`

The residual connection followed by layer normalization.

Defined in Section 11.7.2

`forward(`*X*, *Y*`)`

Defines the computation performed at every call.

Should be overridden by all subclasses.

---

**Note:** Although the recipe for forward pass needs to be defined within this function, one should call the `Module` (page 1113) instance afterwards instead of this since the former takes care of running the registered hooks while the latter silently ignores them.

---

`training: bool`

`class d2l.torch.AttentionDecoder`

Bases: `Decoder` (page 1110)

The base attention-based decoder interface.

Defined in Section 11.4

`property attention_weights`

`training: bool`

`class d2l.torch.Classifier(`*plot_train_per_epoch=2*, *plot_valid_per_epoch=1*`)`

Bases: `Module` (page 1113)

The base class of classification models.

Defined in Section 4.3

`accuracy(`*Y_hat*, *Y*, *averaged=True*`)`

Compute the number of correct predictions.

Defined in Section 4.3

`layer_summary(`*X_shape*`)`

Defined in Section 7.6

`loss(`*Y_hat*, *Y*, *averaged=True*`)`

Defined in Section 4.5

`training: bool`

`validation_step(`*batch*`)`

`class d2l.torch.DataModule(`*root='../data'*, *num_workers=4*`)`

Bases: `HyperParameters` (page 1112)

The base class of data.

Defined in Section 3.2.2

`get_dataloader(`*train*`)`

`get_tensorloader(`*tensors*, *train*, *indices=slice(0, None, None)*`)`

Defined in Section 3.3

`train_dataloader()`

`val_dataloader()`

`class d2l.torch.Decoder`

Bases: `Module`

The base decoder interface for the encoder–decoder architecture.

Defined in Section 10.6

`forward(`*X*, *state*`)`

Defines the computation performed at every call.

Should be overridden by all subclasses.

---

**Note:** Although the recipe for forward pass needs to be defined within this function, one should call the `Module` (page 1113) instance afterwards instead of this since the former takes care of running the registered hooks while the latter silently ignores them.

---

`init_state(`*enc_all_outputs*, **args*`)`

`training: bool`

`class d2l.torch.DotProductAttention(`*dropout*`)`

Bases: `Module`

Scaled dot product attention.

Defined in Section 11.3.2

`forward(`*queries*, *keys*, *values*, *valid_lens=None*`)`

Defines the computation performed at every call.

Should be overridden by all subclasses.

---

**Note:** Although the recipe for forward pass needs to be defined within this function, one should call the `Module` (page 1113) instance afterwards instead of this since the former takes care of running the registered hooks while the latter silently ignores them.

---

`training: bool`

`class d2l.torch.Encoder`

Bases: `Module`

The base encoder interface for the encoder–decoder architecture.

Defined in Section 10.6

`forward(`*X*, **args*`)`

Defines the computation performed at every call.

Should be overridden by all subclasses.

---

**Note:** Although the recipe for forward pass needs to be defined within this function, one should call the `Module` (page 1113) instance afterwards instead of this since the former takes care of running the registered hooks while the latter silently ignores them.

---

`training: bool`

`class d2l.torch.EncoderDecoder(`*encoder*, *decoder*`)`

Bases: `Classifier` (page 1109)

The base class for the encoder–decoder architecture.

Defined in Section 10.6

`forward(`*enc_X*, *dec_X*, **args*`)`

Defines the computation performed at every call.

Should be overridden by all subclasses.

---

**Note:** Although the recipe for forward pass needs to be defined within this function, one should call the `Module` (page 1113) instance afterwards instead of this since the former takes care of running the registered hooks while the latter silently ignores them.

---

`predict_step`(*batch*, *device*, *num_steps*, *save_attention_weights=False*)

Defined in Section 10.7.6

`training: bool`

`class d2l.torch.FashionMNIST`(*batch_size=64*, *resize=(28, 28)*)

Bases: `DataModule` (page 1110)

The Fashion-MNIST dataset.

Defined in Section 4.2

`get_dataloader`(*train*)

Defined in Section 4.2

`text_labels`(*indices*)

Return text labels.

Defined in Section 4.2

`visualize`(*batch*, *nrows=1*, *ncols=8*, *labels=[]*)

Defined in Section 4.2

`class d2l.torch.GRU`(*num_inputs*, *num_hiddens*, *num_layers*, *dropout=0*)

Bases: `RNN` (page 1117)

The multilayer GRU model.

Defined in Section 10.3

`training: bool`

`class d2l.torch.HyperParameters`

Bases: `object`

The base class of hyperparameters.

`save_hyperparameters`(*ignore=[]*)

Save function arguments into class attributes.

Defined in Section B.7

`class d2l.torch.LeNet`(*lr=0.1*, *num_classes=10*)

Bases: `Classifier` (page 1109)

The LeNet-5 model.

Defined in Section 7.6

`training: bool`

`class d2l.torch.LinearRegression(`*lr*`)`

Bases: `Module` (page 1113)

The linear regression model implemented with high-level APIs.

Defined in Section 3.5

`configure_optimizers()`

Defined in Section 3.5

`forward(`*X*`)`

Defined in Section 3.5

`get_w_b()`

Defined in Section 3.5

`loss(`*y_hat*, *y*`)`

Defined in Section 3.5

`training: bool`

`class d2l.torch.LinearRegressionScratch(`*num_inputs*, *lr*, *sigma=0.01*`)`

Bases: `Module` (page 1113)

The linear regression model implemented from scratch.

Defined in Section 3.4

`configure_optimizers()`

Defined in Section 3.4

`forward(`*X*`)`

Defined in Section 3.4

`loss(`*y_hat*, *y*`)`

Defined in Section 3.4

`training: bool`

`class d2l.torch.Module(`*plot_train_per_epoch=2*, *plot_valid_per_epoch=1*`)`

Bases: `Module`, `HyperParameters` (page 1112)

The base class of models.

Defined in Section 3.2

`apply_init(`*inputs*, *init=None*`)`

Defined in Section 6.4

`configure_optimizers()`

Defined in Section 4.3

`forward`(*X*)

Defines the computation performed at every call.

Should be overridden by all subclasses.

---

**Note:** Although the recipe for forward pass needs to be defined within this function, one should call the `Module` (page 1113) instance afterwards instead of this since the former takes care of running the registered hooks while the latter silently ignores them.

---

`loss`(*y_hat*, *y*)

`plot`(*key*, *value*, *train*)

Plot a point in animation.

`training: bool`

`training_step`(*batch*)

`validation_step`(*batch*)

`class d2l.torch.MTFraEng`(*batch_size*, *num_steps=9*, *num_train=512*, *num_val=128*)

Bases: `DataModule` (page 1110)

The English-French dataset.

Defined in Section 10.5

`build`(*src_sentences*, *tgt_sentences*)

Defined in Section 10.5.3

`get_dataloader`(*train*)

Defined in Section 10.5.3

`class d2l.torch.MultiHeadAttention`(*num_hiddens*, *num_heads*, *dropout*, *bias=False*, ***kwargs*)

Bases: `Module` (page 1113)

Multi-head attention.

Defined in Section 11.5

`forward`(*queries*, *keys*, *values*, *valid_lens*)

Defines the computation performed at every call.

Should be overridden by all subclasses.

---

**Note:** Although the recipe for forward pass needs to be defined within this function,

one should call the `Module` (page 1113) instance afterwards instead of this since the former takes care of running the registered hooks while the latter silently ignores them.

`training: bool`

`transpose_output(X)`

Reverse the operation of transpose_qkv.

Defined in Section 11.5

`transpose_qkv(X)`

Transposition for parallel computation of multiple attention heads.

Defined in Section 11.5

`class d2l.torch.PositionalEncoding(num_hiddens, dropout, max_len=1000)`

Bases: `Module`

Positional encoding.

Defined in Section 11.6

`forward(X)`

Defines the computation performed at every call.

Should be overridden by all subclasses.

**Note:** Although the recipe for forward pass needs to be defined within this function, one should call the `Module` (page 1113) instance afterwards instead of this since the former takes care of running the registered hooks while the latter silently ignores them.

`training: bool`

`class d2l.torch.PositionWiseFFN(ffn_num_hiddens, ffn_num_outputs)`

Bases: `Module`

The positionwise feed-forward network.

Defined in Section 11.7

`forward(X)`

Defines the computation performed at every call.

Should be overridden by all subclasses.

**Note:** Although the recipe for forward pass needs to be defined within this function,

one should call the `Module` (page 1113) instance afterwards instead of this since the former takes care of running the registered hooks while the latter silently ignores them.

---

`training: bool`

`class d2l.torch.ProgressBoard(`*xlabel=None*, *ylabel=None*, *xlim=None*, *ylim=None*, *xscale='linear'*, *yscale='linear'*, *ls=['-', '--', '-.', ':']*, *colors=['C0', 'C1', 'C2', 'C3']*, *fig=None*, *axes=None*, *figsize=(3.5, 2.5)*, *display=True*`)`

Bases: `HyperParameters` (page 1112)

The board that plots data points in animation.

Defined in Section 3.2

`draw(`*x*, *y*, *label*, *every_n=1*`)`

Defined in Section B.7

`class d2l.torch.Residual(`*num_channels*, *use_1x1conv=False*, *strides=1*`)`

Bases: `Module`

The Residual block of ResNet models.

Defined in Section 8.6

`forward(`*X*`)`

Defines the computation performed at every call.

Should be overridden by all subclasses.

---

**Note:** Although the recipe for forward pass needs to be defined within this function, one should call the `Module` (page 1113) instance afterwards instead of this since the former takes care of running the registered hooks while the latter silently ignores them.

---

`training: bool`

`class d2l.torch.ResNeXtBlock(`*num_channels*, *groups*, *bot_mul*, *use_1x1conv=False*, *strides=1*`)`

Bases: `Module`

The ResNeXt block.

Defined in Section 8.6.2

`forward(`*X*`)`

Defines the computation performed at every call.

Should be overridden by all subclasses.

---

**Note:** Although the recipe for forward pass needs to be defined within this function, one should call the `Module` (page 1113) instance afterwards instead of this since the former takes care of running the registered hooks while the latter silently ignores them.

---

`training: bool`

`class d2l.torch.RNN(`*num_inputs*, *num_hiddens*`)`

Bases: `Module` (page 1113)

The RNN model implemented with high-level APIs.

Defined in Section 9.6

`forward(`*inputs*, *H=None*`)`

Defines the computation performed at every call.

Should be overridden by all subclasses.

---

**Note:** Although the recipe for forward pass needs to be defined within this function, one should call the `Module` (page 1113) instance afterwards instead of this since the former takes care of running the registered hooks while the latter silently ignores them.

---

`training: bool`

`class d2l.torch.RNNLM(`*rnn*, *vocab_size*, *lr=0.01*`)`

Bases: `RNNLMScratch` (page 1117)

The RNN-based language model implemented with high-level APIs.

Defined in Section 9.6

`init_params()`

`output_layer(`*hiddens*`)`

Defined in Section 9.5

`training: bool`

`class d2l.torch.RNNLMScratch(`*rnn*, *vocab_size*, *lr=0.01*`)`

Bases: `Classifier` (page 1109)

The RNN-based language model implemented from scratch.

Defined in Section 9.5

`forward(`*X*, *state=None*`)`

Defined in Section 9.5

`init_params()`

`one_hot(`*X*`)`

Defined in Section 9.5

`output_layer(`*rnn_outputs*`)`

Defined in Section 9.5

`predict(`*prefix*, *num_preds*, *vocab*, *device=None*`)`

Defined in Section 9.5

`training: bool`

`training_step(`*batch*`)`

`validation_step(`*batch*`)`

`class d2l.torch.RNNScratch(`*num_inputs*, *num_hiddens*, *sigma=0.01*`)`

Bases: `Module` (page 1113)

The RNN model implemented from scratch.

Defined in Section 9.5

`forward(`*inputs*, *state=None*`)`

Defined in Section 9.5

`training: bool`

`class d2l.torch.Seq2Seq(`*encoder*, *decoder*, *tgt_pad*, *lr*`)`

Bases: `EncoderDecoder` (page 1111)

The RNN encoder–decoder for sequence to sequence learning.

Defined in Section 10.7.3

`configure_optimizers()`

Defined in Section 4.3

`training: bool`

`validation_step(`*batch*`)`

`class d2l.torch.Seq2SeqEncoder(`*vocab_size*, *embed_size*, *num_hiddens*, *num_layers*, *dropout=0*`)`

Bases: `Encoder` (page 1111)

The RNN encoder for sequence-to-sequence learning.

Defined in Section 10.7

`forward(`*X*, **args*`)`

Defines the computation performed at every call.

Should be overridden by all subclasses.

---

**Note:** Although the recipe for forward pass needs to be defined within this function, one should call the `Module` (page 1113) instance afterwards instead of this since the former takes care of running the registered hooks while the latter silently ignores them.

---

`training: bool`

`class d2l.torch.SGD(`*params*, *lr*`)`

Bases: `HyperParameters` (page 1112)

Minibatch stochastic gradient descent.

Defined in Section 3.4

`step()`

`zero_grad()`

`class d2l.torch.SoftmaxRegression(`*num_outputs*, *lr*`)`

Bases: `Classifier` (page 1109)

The softmax regression model.

Defined in Section 4.5

`forward(`*X*`)`

Defines the computation performed at every call.

Should be overridden by all subclasses.

---

**Note:** Although the recipe for forward pass needs to be defined within this function, one should call the `Module` (page 1113) instance afterwards instead of this since the former takes care of running the registered hooks while the latter silently ignores them.

---

`training: bool`

`class d2l.torch.SyntheticRegressionData(`*w*, *b*, *noise=0.01*, *num_train=1000*, *num_val=1000*, *batch_size=32*`)`

Bases: `DataModule` (page 1110)

Synthetic data for linear regression.

Defined in Section 3.3

`get_dataloader(`*train*`)`

Defined in Section 3.3

`class d2l.torch.TimeMachine(`*batch_size*, *num_steps*, *num_train=10000*, *num_val=5000*`)`

Bases: `DataModule` (page 1110)

The Time Machine dataset.

Defined in Section 9.2

`build(`*raw_text*, *vocab=None*`)`

Defined in Section 9.2

`get_dataloader(`*train*`)`

Defined in Section 9.3.3

`class d2l.torch.Trainer(`*max_epochs*, *num_gpus=0*, *gradient_clip_val=0*`)`

Bases: `HyperParameters` (page 1112)

The base class for training models with data.

Defined in Section 3.2.2

`clip_gradients(`*grad_clip_val*, *model*`)`

Defined in Section 9.5

`fit(`*model*, *data*`)`

`fit_epoch()`

Defined in Section 3.4

`prepare_batch(`*batch*`)`

Defined in Section 6.7

`prepare_data(`*data*`)`

`prepare_model(`*model*`)`

Defined in Section 6.7

`class d2l.torch.TransformerEncoder(`*vocab_size*, *num_hiddens*, *ffn_num_hiddens*, *num_heads*, *num_blks*, *dropout*, *use_bias=False*`)`

Bases: `Encoder` (page 1111)

The Transformer encoder.

Defined in Section 11.7.4

`forward(`*X*, *valid_lens*`)`

Defines the computation performed at every call.

Should be overridden by all subclasses.

---

**Note:** Although the recipe for forward pass needs to be defined within this function, one should call the `Module` (page 1113) instance afterwards instead of this since the former takes care of running the registered hooks while the latter silently ignores them.

---

`training: bool`

`class d2l.torch.TransformerEncoderBlock(`*num_hiddens*, *ffn_num_hiddens*, *num_heads*, *dropout*, *use_bias=False*`)`

Bases: `Module`

The Transformer encoder block.

Defined in Section 11.7.2

`forward(`*X*, *valid_lens*`)`

Defines the computation performed at every call.

Should be overridden by all subclasses.

---

**Note:** Although the recipe for forward pass needs to be defined within this function, one should call the `Module` (page 1113) instance afterwards instead of this since the former takes care of running the registered hooks while the latter silently ignores them.

---

`training: bool`

`class d2l.torch.Vocab(`*tokens=[]*, *min_freq=0*, *reserved_tokens=[]*`)`

Bases: `object`

Vocabulary for text.

`to_tokens(`*indices*`)`

`property unk`

### B.8.2 Functions

`d2l.torch.add_to_class(`*Class*`)`

Register functions as methods in created class.

Defined in Section 3.2

`d2l.torch.bleu(`*pred_seq*, *label_seq*, *k*`)`

Compute the BLEU.

Defined in Section 10.7.6

`d2l.torch.check_len(`*a*, *n*`)`

Check the length of a list.

Defined in Section 9.5

`d2l.torch.check_shape(`*a*, *shape*`)`

Check the shape of a tensor.

Defined in Section 9.5

`d2l.torch.corr2d(`*X*, *K*`)`

Compute 2D cross-correlation.

Defined in Section 7.2

`d2l.torch.cpu()`

Get the CPU device.

Defined in Section 6.7

`d2l.torch.gpu(`*i=0*`)`

Get a GPU device.

Defined in Section 6.7

`d2l.torch.init_cnn(`*module*`)`

Initialize weights for CNNs.

Defined in Section 7.6

`d2l.torch.init_seq2seq(`*module*`)`

Initialize weights for sequence-to-sequence learning.

Defined in Section 10.7

`d2l.torch.masked_softmax(`*X*, *valid_lens*`)`

Perform softmax operation by masking elements on the last axis.

Defined in Section 11.3

`d2l.torch.num_gpus()`

Get the number of available GPUs.

Defined in Section 6.7

`d2l.torch.plot(`*X*, *Y=None*, *xlabel=None*, *ylabel=None*, *legend=[]*, *xlim=None*, *ylim=None*, *xscale='linear'*, *yscale='linear'*, *fmts=('-', 'm--', 'g-.', 'r:')*, *figsize=(3.5, 2.5)*, *axes=None*`)`

Plot data points.

Defined in Section 2.4

`d2l.torch.set_axes(`*axes*, *xlabel*, *ylabel*, *xlim*, *ylim*, *xscale*, *yscale*, *legend*`)`

Set the axes for matplotlib.

Defined in Section 2.4

`d2l.torch.set_figsize(`*figsize=(3.5, 2.5)*`)`

Set the figure size for matplotlib.

Defined in Section 2.4

`d2l.torch.show_heatmaps(`*matrices*, *xlabel*, *ylabel*, *titles=None*, *figsize=(2.5, 2.5)*, *cmap='Reds'*`)`

Show heatmaps of matrices.

Defined in Section 11.1

`d2l.torch.show_list_len_pair_hist(`*legend*, *xlabel*, *ylabel*, *xlist*, *ylist*`)`

Plot the histogram for list length pairs.

Defined in Section 10.5

`d2l.torch.try_all_gpus()`

Return all available GPUs, or [cpu(),] if no GPU exists.

Defined in Section 6.7

`d2l.torch.try_gpu(`*i=0*`)`

Return gpu(i) if exists, otherwise return cpu().

Defined in Section 6.7

`d2l.torch.use_svg_display()`

Use the svg format to display a plot in Jupyter.

Defined in Section 2.4